\documentclass[11pt,letterpaper]{article}
\usepackage[margin=1in]{geometry}
\usepackage[T1]{fontenc}
\usepackage{lmodern}
\usepackage{microtype}
\usepackage{amsmath,amssymb,booktabs,graphicx}
\usepackage{adjustbox,xurl}
\usepackage[numbers,sort&compress]{natbib}
\usepackage[font=small,labelfont=bf]{caption}
\usepackage[hidelinks,unicode]{hyperref}
\usepackage[nameinlink,noabbrev,capitalise]{cleveref}

\newcommand{\PreprintAuthors}{%
  Shengwei Zhang\textsuperscript{1}\quad Shenkang Qian\textsuperscript{2}\\[0.25em]
  Haoda Dai\textsuperscript{1}\quad Yuheng Song\textsuperscript{1}\quad Boming Zhang\textsuperscript{1}\\[0.25em]
  \normalsize\textsuperscript{1}University of Pennsylvania\quad
  \textsuperscript{2}University College London%
}
\author{\PreprintAuthors}
\date{}
\input{numbers}
\makeatletter
\expandafter\def\csname val@rev.desc.d1.sspro.p1.C.center-d1.delta-tau.ci90.0\endcsname{-0.062}
\expandafter\def\csname cnt@rev.desc.d1.sspro.p1.C.center-d1.delta-tau.ci90.0\endcsname{0}
\expandafter\def\csname pct@rev.desc.d1.sspro.p1.C.center-d1.delta-tau.ci90.0\endcsname{$-$6.2}
\expandafter\def\csname ppt@rev.desc.d1.sspro.p1.C.center-d1.delta-tau.ci90.0\endcsname{$-$6.2}
\expandafter\def\csname val@rev.desc.d1.sspro.p1.C.center-d1.delta-tau.ci90.1\endcsname{0.104}
\expandafter\def\csname cnt@rev.desc.d1.sspro.p1.C.center-d1.delta-tau.ci90.1\endcsname{0}
\expandafter\def\csname pct@rev.desc.d1.sspro.p1.C.center-d1.delta-tau.ci90.1\endcsname{10.4}
\expandafter\def\csname ppt@rev.desc.d1.sspro.p1.C.center-d1.delta-tau.ci90.1\endcsname{$+$10.4}
\expandafter\def\csname val@rev.desc.d1.sspro.p1.C.center-d1.delta-tau.ci95-excludes-0\endcsname{0}
\expandafter\def\csname cnt@rev.desc.d1.sspro.p1.C.center-d1.delta-tau.ci95-excludes-0\endcsname{0}
\expandafter\def\csname pct@rev.desc.d1.sspro.p1.C.center-d1.delta-tau.ci95-excludes-0\endcsname{0.0}
\expandafter\def\csname ppt@rev.desc.d1.sspro.p1.C.center-d1.delta-tau.ci95-excludes-0\endcsname{0.0}
\expandafter\def\csname val@rev.desc.d1.sspro.p1.C.center-d1.delta-tau.ci95.0\endcsname{-0.083}
\expandafter\def\csname cnt@rev.desc.d1.sspro.p1.C.center-d1.delta-tau.ci95.0\endcsname{0}
\expandafter\def\csname pct@rev.desc.d1.sspro.p1.C.center-d1.delta-tau.ci95.0\endcsname{$-$8.3}
\expandafter\def\csname ppt@rev.desc.d1.sspro.p1.C.center-d1.delta-tau.ci95.0\endcsname{$-$8.3}
\expandafter\def\csname val@rev.desc.d1.sspro.p1.C.center-d1.delta-tau.ci95.1\endcsname{0.125}
\expandafter\def\csname cnt@rev.desc.d1.sspro.p1.C.center-d1.delta-tau.ci95.1\endcsname{0}
\expandafter\def\csname pct@rev.desc.d1.sspro.p1.C.center-d1.delta-tau.ci95.1\endcsname{12.5}
\expandafter\def\csname ppt@rev.desc.d1.sspro.p1.C.center-d1.delta-tau.ci95.1\endcsname{$+$12.5}
\expandafter\def\csname val@rev.desc.d1.sspro.p1.C.center-d1.delta-tau.mean\endcsname{0.021}
\expandafter\def\csname cnt@rev.desc.d1.sspro.p1.C.center-d1.delta-tau.mean\endcsname{0}
\expandafter\def\csname pct@rev.desc.d1.sspro.p1.C.center-d1.delta-tau.mean\endcsname{2.1}
\expandafter\def\csname ppt@rev.desc.d1.sspro.p1.C.center-d1.delta-tau.mean\endcsname{$+$2.1}
\expandafter\def\csname val@rev.desc.d1.sspro.p1.C.center-d1.delta-tau.n-apps\endcsname{6}
\expandafter\def\csname cnt@rev.desc.d1.sspro.p1.C.center-d1.delta-tau.n-apps\endcsname{6}
\expandafter\def\csname pct@rev.desc.d1.sspro.p1.C.center-d1.delta-tau.n-apps\endcsname{600.0}
\expandafter\def\csname ppt@rev.desc.d1.sspro.p1.C.center-d1.delta-tau.n-apps\endcsname{$+$600.0}
\expandafter\def\csname val@rev.desc.d1.sspro.p1.C.center-d1.delta-tau.tost1.ci90-inside\endcsname{0}
\expandafter\def\csname cnt@rev.desc.d1.sspro.p1.C.center-d1.delta-tau.tost1.ci90-inside\endcsname{0}
\expandafter\def\csname pct@rev.desc.d1.sspro.p1.C.center-d1.delta-tau.tost1.ci90-inside\endcsname{0.0}
\expandafter\def\csname ppt@rev.desc.d1.sspro.p1.C.center-d1.delta-tau.tost1.ci90-inside\endcsname{0.0}
\expandafter\def\csname val@rev.desc.d1.sspro.p1.C.center-d1.delta-tau.tost1.equivalent\endcsname{0}
\expandafter\def\csname cnt@rev.desc.d1.sspro.p1.C.center-d1.delta-tau.tost1.equivalent\endcsname{0}
\expandafter\def\csname pct@rev.desc.d1.sspro.p1.C.center-d1.delta-tau.tost1.equivalent\endcsname{0.0}
\expandafter\def\csname ppt@rev.desc.d1.sspro.p1.C.center-d1.delta-tau.tost1.equivalent\endcsname{0.0}
\expandafter\def\csname val@rev.desc.d1.sspro.p1.C.center-d1.delta-tau.tost1.p\endcsname{0.562}
\expandafter\def\csname cnt@rev.desc.d1.sspro.p1.C.center-d1.delta-tau.tost1.p\endcsname{1}
\expandafter\def\csname pct@rev.desc.d1.sspro.p1.C.center-d1.delta-tau.tost1.p\endcsname{56.2}
\expandafter\def\csname ppt@rev.desc.d1.sspro.p1.C.center-d1.delta-tau.tost1.p\endcsname{$+$56.2}
\expandafter\def\csname val@rev.desc.d1.sspro.p1.C.center-d1.delta-tau.tost5.ci90-inside\endcsname{0}
\expandafter\def\csname cnt@rev.desc.d1.sspro.p1.C.center-d1.delta-tau.tost5.ci90-inside\endcsname{0}
\expandafter\def\csname pct@rev.desc.d1.sspro.p1.C.center-d1.delta-tau.tost5.ci90-inside\endcsname{0.0}
\expandafter\def\csname ppt@rev.desc.d1.sspro.p1.C.center-d1.delta-tau.tost5.ci90-inside\endcsname{0.0}
\expandafter\def\csname val@rev.desc.d1.sspro.p1.C.center-d1.delta-tau.tost5.equivalent\endcsname{0}
\expandafter\def\csname cnt@rev.desc.d1.sspro.p1.C.center-d1.delta-tau.tost5.equivalent\endcsname{0}
\expandafter\def\csname pct@rev.desc.d1.sspro.p1.C.center-d1.delta-tau.tost5.equivalent\endcsname{0.0}
\expandafter\def\csname ppt@rev.desc.d1.sspro.p1.C.center-d1.delta-tau.tost5.equivalent\endcsname{0.0}
\expandafter\def\csname val@rev.desc.d1.sspro.p1.C.center-d1.delta-tau.tost5.p\endcsname{0.269}
\expandafter\def\csname cnt@rev.desc.d1.sspro.p1.C.center-d1.delta-tau.tost5.p\endcsname{0}
\expandafter\def\csname pct@rev.desc.d1.sspro.p1.C.center-d1.delta-tau.tost5.p\endcsname{26.9}
\expandafter\def\csname ppt@rev.desc.d1.sspro.p1.C.center-d1.delta-tau.tost5.p\endcsname{$+$26.9}
\expandafter\def\csname val@rev.desc.d1.sspro.p1.C.center-d1.material-flip\endcsname{0}
\expandafter\def\csname cnt@rev.desc.d1.sspro.p1.C.center-d1.material-flip\endcsname{0}
\expandafter\def\csname pct@rev.desc.d1.sspro.p1.C.center-d1.material-flip\endcsname{0.0}
\expandafter\def\csname ppt@rev.desc.d1.sspro.p1.C.center-d1.material-flip\endcsname{0.0}
\expandafter\def\csname val@rev.desc.d1.sspro.p1.C.center-d1.n-apps\endcsname{6}
\expandafter\def\csname cnt@rev.desc.d1.sspro.p1.C.center-d1.n-apps\endcsname{6}
\expandafter\def\csname pct@rev.desc.d1.sspro.p1.C.center-d1.n-apps\endcsname{600.0}
\expandafter\def\csname ppt@rev.desc.d1.sspro.p1.C.center-d1.n-apps\endcsname{$+$600.0}
\expandafter\def\csname val@rev.desc.d1.sspro.p1.C.center-d1.n-tasks\endcsname{48}
\expandafter\def\csname cnt@rev.desc.d1.sspro.p1.C.center-d1.n-tasks\endcsname{48}
\expandafter\def\csname pct@rev.desc.d1.sspro.p1.C.center-d1.n-tasks\endcsname{4800.0}
\expandafter\def\csname ppt@rev.desc.d1.sspro.p1.C.center-d1.n-tasks\endcsname{$+$4800.0}
\expandafter\def\csname val@rev.desc.d1.sspro.p1.C.center-d1.sign-differs\endcsname{0}
\expandafter\def\csname cnt@rev.desc.d1.sspro.p1.C.center-d1.sign-differs\endcsname{0}
\expandafter\def\csname pct@rev.desc.d1.sspro.p1.C.center-d1.sign-differs\endcsname{0.0}
\expandafter\def\csname ppt@rev.desc.d1.sspro.p1.C.center-d1.sign-differs\endcsname{0.0}
\expandafter\def\csname val@rev.desc.d1.sspro.p1.C.center-d1.tau-source.ci90.0\endcsname{0.042}
\expandafter\def\csname cnt@rev.desc.d1.sspro.p1.C.center-d1.tau-source.ci90.0\endcsname{0}
\expandafter\def\csname pct@rev.desc.d1.sspro.p1.C.center-d1.tau-source.ci90.0\endcsname{4.2}
\expandafter\def\csname ppt@rev.desc.d1.sspro.p1.C.center-d1.tau-source.ci90.0\endcsname{$+$4.2}
\expandafter\def\csname val@rev.desc.d1.sspro.p1.C.center-d1.tau-source.ci90.1\endcsname{0.271}
\expandafter\def\csname cnt@rev.desc.d1.sspro.p1.C.center-d1.tau-source.ci90.1\endcsname{0}
\expandafter\def\csname pct@rev.desc.d1.sspro.p1.C.center-d1.tau-source.ci90.1\endcsname{27.1}
\expandafter\def\csname ppt@rev.desc.d1.sspro.p1.C.center-d1.tau-source.ci90.1\endcsname{$+$27.1}
\expandafter\def\csname val@rev.desc.d1.sspro.p1.C.center-d1.tau-source.ci95-excludes-0\endcsname{1}
\expandafter\def\csname cnt@rev.desc.d1.sspro.p1.C.center-d1.tau-source.ci95-excludes-0\endcsname{1}
\expandafter\def\csname pct@rev.desc.d1.sspro.p1.C.center-d1.tau-source.ci95-excludes-0\endcsname{100.0}
\expandafter\def\csname ppt@rev.desc.d1.sspro.p1.C.center-d1.tau-source.ci95-excludes-0\endcsname{$+$100.0}
\expandafter\def\csname val@rev.desc.d1.sspro.p1.C.center-d1.tau-source.ci95.0\endcsname{0.042}
\expandafter\def\csname cnt@rev.desc.d1.sspro.p1.C.center-d1.tau-source.ci95.0\endcsname{0}
\expandafter\def\csname pct@rev.desc.d1.sspro.p1.C.center-d1.tau-source.ci95.0\endcsname{4.2}
\expandafter\def\csname ppt@rev.desc.d1.sspro.p1.C.center-d1.tau-source.ci95.0\endcsname{$+$4.2}
\expandafter\def\csname val@rev.desc.d1.sspro.p1.C.center-d1.tau-source.ci95.1\endcsname{0.292}
\expandafter\def\csname cnt@rev.desc.d1.sspro.p1.C.center-d1.tau-source.ci95.1\endcsname{0}
\expandafter\def\csname pct@rev.desc.d1.sspro.p1.C.center-d1.tau-source.ci95.1\endcsname{29.2}
\expandafter\def\csname ppt@rev.desc.d1.sspro.p1.C.center-d1.tau-source.ci95.1\endcsname{$+$29.2}
\expandafter\def\csname val@rev.desc.d1.sspro.p1.C.center-d1.tau-source.mean\endcsname{0.146}
\expandafter\def\csname cnt@rev.desc.d1.sspro.p1.C.center-d1.tau-source.mean\endcsname{0}
\expandafter\def\csname pct@rev.desc.d1.sspro.p1.C.center-d1.tau-source.mean\endcsname{14.6}
\expandafter\def\csname ppt@rev.desc.d1.sspro.p1.C.center-d1.tau-source.mean\endcsname{$+$14.6}
\expandafter\def\csname val@rev.desc.d1.sspro.p1.C.center-d1.tau-source.n-apps\endcsname{6}
\expandafter\def\csname cnt@rev.desc.d1.sspro.p1.C.center-d1.tau-source.n-apps\endcsname{6}
\expandafter\def\csname pct@rev.desc.d1.sspro.p1.C.center-d1.tau-source.n-apps\endcsname{600.0}
\expandafter\def\csname ppt@rev.desc.d1.sspro.p1.C.center-d1.tau-source.n-apps\endcsname{$+$600.0}
\expandafter\def\csname val@rev.desc.d1.sspro.p1.C.center-d1.tau-source.tost1.ci90-inside\endcsname{0}
\expandafter\def\csname cnt@rev.desc.d1.sspro.p1.C.center-d1.tau-source.tost1.ci90-inside\endcsname{0}
\expandafter\def\csname pct@rev.desc.d1.sspro.p1.C.center-d1.tau-source.tost1.ci90-inside\endcsname{0.0}
\expandafter\def\csname ppt@rev.desc.d1.sspro.p1.C.center-d1.tau-source.tost1.ci90-inside\endcsname{0.0}
\expandafter\def\csname val@rev.desc.d1.sspro.p1.C.center-d1.tau-source.tost1.equivalent\endcsname{0}
\expandafter\def\csname cnt@rev.desc.d1.sspro.p1.C.center-d1.tau-source.tost1.equivalent\endcsname{0}
\expandafter\def\csname pct@rev.desc.d1.sspro.p1.C.center-d1.tau-source.tost1.equivalent\endcsname{0.0}
\expandafter\def\csname ppt@rev.desc.d1.sspro.p1.C.center-d1.tau-source.tost1.equivalent\endcsname{0.0}
\expandafter\def\csname val@rev.desc.d1.sspro.p1.C.center-d1.tau-source.tost1.p\endcsname{0.999}
\expandafter\def\csname cnt@rev.desc.d1.sspro.p1.C.center-d1.tau-source.tost1.p\endcsname{1}
\expandafter\def\csname pct@rev.desc.d1.sspro.p1.C.center-d1.tau-source.tost1.p\endcsname{99.9}
\expandafter\def\csname ppt@rev.desc.d1.sspro.p1.C.center-d1.tau-source.tost1.p\endcsname{$+$99.9}
\expandafter\def\csname val@rev.desc.d1.sspro.p1.C.center-d1.tau-source.tost5.ci90-inside\endcsname{0}
\expandafter\def\csname cnt@rev.desc.d1.sspro.p1.C.center-d1.tau-source.tost5.ci90-inside\endcsname{0}
\expandafter\def\csname pct@rev.desc.d1.sspro.p1.C.center-d1.tau-source.tost5.ci90-inside\endcsname{0.0}
\expandafter\def\csname ppt@rev.desc.d1.sspro.p1.C.center-d1.tau-source.tost5.ci90-inside\endcsname{0.0}
\expandafter\def\csname val@rev.desc.d1.sspro.p1.C.center-d1.tau-source.tost5.equivalent\endcsname{0}
\expandafter\def\csname cnt@rev.desc.d1.sspro.p1.C.center-d1.tau-source.tost5.equivalent\endcsname{0}
\expandafter\def\csname pct@rev.desc.d1.sspro.p1.C.center-d1.tau-source.tost5.equivalent\endcsname{0.0}
\expandafter\def\csname ppt@rev.desc.d1.sspro.p1.C.center-d1.tau-source.tost5.equivalent\endcsname{0.0}
\expandafter\def\csname val@rev.desc.d1.sspro.p1.C.center-d1.tau-source.tost5.p\endcsname{0.938}
\expandafter\def\csname cnt@rev.desc.d1.sspro.p1.C.center-d1.tau-source.tost5.p\endcsname{1}
\expandafter\def\csname pct@rev.desc.d1.sspro.p1.C.center-d1.tau-source.tost5.p\endcsname{93.8}
\expandafter\def\csname ppt@rev.desc.d1.sspro.p1.C.center-d1.tau-source.tost5.p\endcsname{$+$93.8}
\expandafter\def\csname val@rev.desc.d1.sspro.p1.C.center-d1.tau-target.ci90.0\endcsname{0.062}
\expandafter\def\csname cnt@rev.desc.d1.sspro.p1.C.center-d1.tau-target.ci90.0\endcsname{0}
\expandafter\def\csname pct@rev.desc.d1.sspro.p1.C.center-d1.tau-target.ci90.0\endcsname{6.2}
\expandafter\def\csname ppt@rev.desc.d1.sspro.p1.C.center-d1.tau-target.ci90.0\endcsname{$+$6.2}
\expandafter\def\csname val@rev.desc.d1.sspro.p1.C.center-d1.tau-target.ci90.1\endcsname{0.271}
\expandafter\def\csname cnt@rev.desc.d1.sspro.p1.C.center-d1.tau-target.ci90.1\endcsname{0}
\expandafter\def\csname pct@rev.desc.d1.sspro.p1.C.center-d1.tau-target.ci90.1\endcsname{27.1}
\expandafter\def\csname ppt@rev.desc.d1.sspro.p1.C.center-d1.tau-target.ci90.1\endcsname{$+$27.1}
\expandafter\def\csname val@rev.desc.d1.sspro.p1.C.center-d1.tau-target.ci95-excludes-0\endcsname{1}
\expandafter\def\csname cnt@rev.desc.d1.sspro.p1.C.center-d1.tau-target.ci95-excludes-0\endcsname{1}
\expandafter\def\csname pct@rev.desc.d1.sspro.p1.C.center-d1.tau-target.ci95-excludes-0\endcsname{100.0}
\expandafter\def\csname ppt@rev.desc.d1.sspro.p1.C.center-d1.tau-target.ci95-excludes-0\endcsname{$+$100.0}
\expandafter\def\csname val@rev.desc.d1.sspro.p1.C.center-d1.tau-target.ci95.0\endcsname{0.042}
\expandafter\def\csname cnt@rev.desc.d1.sspro.p1.C.center-d1.tau-target.ci95.0\endcsname{0}
\expandafter\def\csname pct@rev.desc.d1.sspro.p1.C.center-d1.tau-target.ci95.0\endcsname{4.2}
\expandafter\def\csname ppt@rev.desc.d1.sspro.p1.C.center-d1.tau-target.ci95.0\endcsname{$+$4.2}
\expandafter\def\csname val@rev.desc.d1.sspro.p1.C.center-d1.tau-target.ci95.1\endcsname{0.292}
\expandafter\def\csname cnt@rev.desc.d1.sspro.p1.C.center-d1.tau-target.ci95.1\endcsname{0}
\expandafter\def\csname pct@rev.desc.d1.sspro.p1.C.center-d1.tau-target.ci95.1\endcsname{29.2}
\expandafter\def\csname ppt@rev.desc.d1.sspro.p1.C.center-d1.tau-target.ci95.1\endcsname{$+$29.2}
\expandafter\def\csname val@rev.desc.d1.sspro.p1.C.center-d1.tau-target.mean\endcsname{0.167}
\expandafter\def\csname cnt@rev.desc.d1.sspro.p1.C.center-d1.tau-target.mean\endcsname{0}
\expandafter\def\csname pct@rev.desc.d1.sspro.p1.C.center-d1.tau-target.mean\endcsname{16.7}
\expandafter\def\csname ppt@rev.desc.d1.sspro.p1.C.center-d1.tau-target.mean\endcsname{$+$16.7}
\expandafter\def\csname val@rev.desc.d1.sspro.p1.C.center-d1.tau-target.n-apps\endcsname{6}
\expandafter\def\csname cnt@rev.desc.d1.sspro.p1.C.center-d1.tau-target.n-apps\endcsname{6}
\expandafter\def\csname pct@rev.desc.d1.sspro.p1.C.center-d1.tau-target.n-apps\endcsname{600.0}
\expandafter\def\csname ppt@rev.desc.d1.sspro.p1.C.center-d1.tau-target.n-apps\endcsname{$+$600.0}
\expandafter\def\csname val@rev.desc.d1.sspro.p1.C.center-d1.tau-target.tost1.ci90-inside\endcsname{0}
\expandafter\def\csname cnt@rev.desc.d1.sspro.p1.C.center-d1.tau-target.tost1.ci90-inside\endcsname{0}
\expandafter\def\csname pct@rev.desc.d1.sspro.p1.C.center-d1.tau-target.tost1.ci90-inside\endcsname{0.0}
\expandafter\def\csname ppt@rev.desc.d1.sspro.p1.C.center-d1.tau-target.tost1.ci90-inside\endcsname{0.0}
\expandafter\def\csname val@rev.desc.d1.sspro.p1.C.center-d1.tau-target.tost1.equivalent\endcsname{0}
\expandafter\def\csname cnt@rev.desc.d1.sspro.p1.C.center-d1.tau-target.tost1.equivalent\endcsname{0}
\expandafter\def\csname pct@rev.desc.d1.sspro.p1.C.center-d1.tau-target.tost1.equivalent\endcsname{0.0}
\expandafter\def\csname ppt@rev.desc.d1.sspro.p1.C.center-d1.tau-target.tost1.equivalent\endcsname{0.0}
\expandafter\def\csname val@rev.desc.d1.sspro.p1.C.center-d1.tau-target.tost1.p\endcsname{0.999}
\expandafter\def\csname cnt@rev.desc.d1.sspro.p1.C.center-d1.tau-target.tost1.p\endcsname{1}
\expandafter\def\csname pct@rev.desc.d1.sspro.p1.C.center-d1.tau-target.tost1.p\endcsname{99.9}
\expandafter\def\csname ppt@rev.desc.d1.sspro.p1.C.center-d1.tau-target.tost1.p\endcsname{$+$99.9}
\expandafter\def\csname val@rev.desc.d1.sspro.p1.C.center-d1.tau-target.tost5.ci90-inside\endcsname{0}
\expandafter\def\csname cnt@rev.desc.d1.sspro.p1.C.center-d1.tau-target.tost5.ci90-inside\endcsname{0}
\expandafter\def\csname pct@rev.desc.d1.sspro.p1.C.center-d1.tau-target.tost5.ci90-inside\endcsname{0.0}
\expandafter\def\csname ppt@rev.desc.d1.sspro.p1.C.center-d1.tau-target.tost5.ci90-inside\endcsname{0.0}
\expandafter\def\csname val@rev.desc.d1.sspro.p1.C.center-d1.tau-target.tost5.equivalent\endcsname{0}
\expandafter\def\csname cnt@rev.desc.d1.sspro.p1.C.center-d1.tau-target.tost5.equivalent\endcsname{0}
\expandafter\def\csname pct@rev.desc.d1.sspro.p1.C.center-d1.tau-target.tost5.equivalent\endcsname{0.0}
\expandafter\def\csname ppt@rev.desc.d1.sspro.p1.C.center-d1.tau-target.tost5.equivalent\endcsname{0.0}
\expandafter\def\csname val@rev.desc.d1.sspro.p1.C.center-d1.tau-target.tost5.p\endcsname{0.971}
\expandafter\def\csname cnt@rev.desc.d1.sspro.p1.C.center-d1.tau-target.tost5.p\endcsname{1}
\expandafter\def\csname pct@rev.desc.d1.sspro.p1.C.center-d1.tau-target.tost5.p\endcsname{97.1}
\expandafter\def\csname ppt@rev.desc.d1.sspro.p1.C.center-d1.tau-target.tost5.p\endcsname{$+$97.1}
\expandafter\def\csname val@rev.desc.d1.sspro.p1.C.center-d1.zero-estimate\endcsname{0}
\expandafter\def\csname cnt@rev.desc.d1.sspro.p1.C.center-d1.zero-estimate\endcsname{0}
\expandafter\def\csname pct@rev.desc.d1.sspro.p1.C.center-d1.zero-estimate\endcsname{0.0}
\expandafter\def\csname ppt@rev.desc.d1.sspro.p1.C.center-d1.zero-estimate\endcsname{0.0}
\expandafter\def\csname val@rev.desc.d1.sspro.p1.C.context-d1.delta-tau.ci90.0\endcsname{-0.167}
\expandafter\def\csname cnt@rev.desc.d1.sspro.p1.C.context-d1.delta-tau.ci90.0\endcsname{0}
\expandafter\def\csname pct@rev.desc.d1.sspro.p1.C.context-d1.delta-tau.ci90.0\endcsname{$-$16.7}
\expandafter\def\csname ppt@rev.desc.d1.sspro.p1.C.context-d1.delta-tau.ci90.0\endcsname{$-$16.7}
\expandafter\def\csname val@rev.desc.d1.sspro.p1.C.context-d1.delta-tau.ci90.1\endcsname{0.021}
\expandafter\def\csname cnt@rev.desc.d1.sspro.p1.C.context-d1.delta-tau.ci90.1\endcsname{0}
\expandafter\def\csname pct@rev.desc.d1.sspro.p1.C.context-d1.delta-tau.ci90.1\endcsname{2.1}
\expandafter\def\csname ppt@rev.desc.d1.sspro.p1.C.context-d1.delta-tau.ci90.1\endcsname{$+$2.1}
\expandafter\def\csname val@rev.desc.d1.sspro.p1.C.context-d1.delta-tau.ci95-excludes-0\endcsname{0}
\expandafter\def\csname cnt@rev.desc.d1.sspro.p1.C.context-d1.delta-tau.ci95-excludes-0\endcsname{0}
\expandafter\def\csname pct@rev.desc.d1.sspro.p1.C.context-d1.delta-tau.ci95-excludes-0\endcsname{0.0}
\expandafter\def\csname ppt@rev.desc.d1.sspro.p1.C.context-d1.delta-tau.ci95-excludes-0\endcsname{0.0}
\expandafter\def\csname val@rev.desc.d1.sspro.p1.C.context-d1.delta-tau.ci95.0\endcsname{-0.167}
\expandafter\def\csname cnt@rev.desc.d1.sspro.p1.C.context-d1.delta-tau.ci95.0\endcsname{0}
\expandafter\def\csname pct@rev.desc.d1.sspro.p1.C.context-d1.delta-tau.ci95.0\endcsname{$-$16.7}
\expandafter\def\csname ppt@rev.desc.d1.sspro.p1.C.context-d1.delta-tau.ci95.0\endcsname{$-$16.7}
\expandafter\def\csname val@rev.desc.d1.sspro.p1.C.context-d1.delta-tau.ci95.1\endcsname{0.042}
\expandafter\def\csname cnt@rev.desc.d1.sspro.p1.C.context-d1.delta-tau.ci95.1\endcsname{0}
\expandafter\def\csname pct@rev.desc.d1.sspro.p1.C.context-d1.delta-tau.ci95.1\endcsname{4.2}
\expandafter\def\csname ppt@rev.desc.d1.sspro.p1.C.context-d1.delta-tau.ci95.1\endcsname{$+$4.2}
\expandafter\def\csname val@rev.desc.d1.sspro.p1.C.context-d1.delta-tau.mean\endcsname{-0.062}
\expandafter\def\csname cnt@rev.desc.d1.sspro.p1.C.context-d1.delta-tau.mean\endcsname{0}
\expandafter\def\csname pct@rev.desc.d1.sspro.p1.C.context-d1.delta-tau.mean\endcsname{$-$6.2}
\expandafter\def\csname ppt@rev.desc.d1.sspro.p1.C.context-d1.delta-tau.mean\endcsname{$-$6.2}
\expandafter\def\csname val@rev.desc.d1.sspro.p1.C.context-d1.delta-tau.n-apps\endcsname{6}
\expandafter\def\csname cnt@rev.desc.d1.sspro.p1.C.context-d1.delta-tau.n-apps\endcsname{6}
\expandafter\def\csname pct@rev.desc.d1.sspro.p1.C.context-d1.delta-tau.n-apps\endcsname{600.0}
\expandafter\def\csname ppt@rev.desc.d1.sspro.p1.C.context-d1.delta-tau.n-apps\endcsname{$+$600.0}
\expandafter\def\csname val@rev.desc.d1.sspro.p1.C.context-d1.delta-tau.tost1.ci90-inside\endcsname{0}
\expandafter\def\csname cnt@rev.desc.d1.sspro.p1.C.context-d1.delta-tau.tost1.ci90-inside\endcsname{0}
\expandafter\def\csname pct@rev.desc.d1.sspro.p1.C.context-d1.delta-tau.tost1.ci90-inside\endcsname{0.0}
\expandafter\def\csname ppt@rev.desc.d1.sspro.p1.C.context-d1.delta-tau.tost1.ci90-inside\endcsname{0.0}
\expandafter\def\csname val@rev.desc.d1.sspro.p1.C.context-d1.delta-tau.tost1.equivalent\endcsname{0}
\expandafter\def\csname cnt@rev.desc.d1.sspro.p1.C.context-d1.delta-tau.tost1.equivalent\endcsname{0}
\expandafter\def\csname pct@rev.desc.d1.sspro.p1.C.context-d1.delta-tau.tost1.equivalent\endcsname{0.0}
\expandafter\def\csname ppt@rev.desc.d1.sspro.p1.C.context-d1.delta-tau.tost1.equivalent\endcsname{0.0}
\expandafter\def\csname val@rev.desc.d1.sspro.p1.C.context-d1.delta-tau.tost1.p\endcsname{0.815}
\expandafter\def\csname cnt@rev.desc.d1.sspro.p1.C.context-d1.delta-tau.tost1.p\endcsname{1}
\expandafter\def\csname pct@rev.desc.d1.sspro.p1.C.context-d1.delta-tau.tost1.p\endcsname{81.5}
\expandafter\def\csname ppt@rev.desc.d1.sspro.p1.C.context-d1.delta-tau.tost1.p\endcsname{$+$81.5}
\expandafter\def\csname val@rev.desc.d1.sspro.p1.C.context-d1.delta-tau.tost5.ci90-inside\endcsname{0}
\expandafter\def\csname cnt@rev.desc.d1.sspro.p1.C.context-d1.delta-tau.tost5.ci90-inside\endcsname{0}
\expandafter\def\csname pct@rev.desc.d1.sspro.p1.C.context-d1.delta-tau.tost5.ci90-inside\endcsname{0.0}
\expandafter\def\csname ppt@rev.desc.d1.sspro.p1.C.context-d1.delta-tau.tost5.ci90-inside\endcsname{0.0}
\expandafter\def\csname val@rev.desc.d1.sspro.p1.C.context-d1.delta-tau.tost5.equivalent\endcsname{0}
\expandafter\def\csname cnt@rev.desc.d1.sspro.p1.C.context-d1.delta-tau.tost5.equivalent\endcsname{0}
\expandafter\def\csname pct@rev.desc.d1.sspro.p1.C.context-d1.delta-tau.tost5.equivalent\endcsname{0.0}
\expandafter\def\csname ppt@rev.desc.d1.sspro.p1.C.context-d1.delta-tau.tost5.equivalent\endcsname{0.0}
\expandafter\def\csname val@rev.desc.d1.sspro.p1.C.context-d1.delta-tau.tost5.p\endcsname{0.551}
\expandafter\def\csname cnt@rev.desc.d1.sspro.p1.C.context-d1.delta-tau.tost5.p\endcsname{1}
\expandafter\def\csname pct@rev.desc.d1.sspro.p1.C.context-d1.delta-tau.tost5.p\endcsname{55.1}
\expandafter\def\csname ppt@rev.desc.d1.sspro.p1.C.context-d1.delta-tau.tost5.p\endcsname{$+$55.1}
\expandafter\def\csname val@rev.desc.d1.sspro.p1.C.context-d1.material-flip\endcsname{0}
\expandafter\def\csname cnt@rev.desc.d1.sspro.p1.C.context-d1.material-flip\endcsname{0}
\expandafter\def\csname pct@rev.desc.d1.sspro.p1.C.context-d1.material-flip\endcsname{0.0}
\expandafter\def\csname ppt@rev.desc.d1.sspro.p1.C.context-d1.material-flip\endcsname{0.0}
\expandafter\def\csname val@rev.desc.d1.sspro.p1.C.context-d1.n-apps\endcsname{6}
\expandafter\def\csname cnt@rev.desc.d1.sspro.p1.C.context-d1.n-apps\endcsname{6}
\expandafter\def\csname pct@rev.desc.d1.sspro.p1.C.context-d1.n-apps\endcsname{600.0}
\expandafter\def\csname ppt@rev.desc.d1.sspro.p1.C.context-d1.n-apps\endcsname{$+$600.0}
\expandafter\def\csname val@rev.desc.d1.sspro.p1.C.context-d1.n-tasks\endcsname{48}
\expandafter\def\csname cnt@rev.desc.d1.sspro.p1.C.context-d1.n-tasks\endcsname{48}
\expandafter\def\csname pct@rev.desc.d1.sspro.p1.C.context-d1.n-tasks\endcsname{4800.0}
\expandafter\def\csname ppt@rev.desc.d1.sspro.p1.C.context-d1.n-tasks\endcsname{$+$4800.0}
\expandafter\def\csname val@rev.desc.d1.sspro.p1.C.context-d1.sign-differs\endcsname{0}
\expandafter\def\csname cnt@rev.desc.d1.sspro.p1.C.context-d1.sign-differs\endcsname{0}
\expandafter\def\csname pct@rev.desc.d1.sspro.p1.C.context-d1.sign-differs\endcsname{0.0}
\expandafter\def\csname ppt@rev.desc.d1.sspro.p1.C.context-d1.sign-differs\endcsname{0.0}
\expandafter\def\csname val@rev.desc.d1.sspro.p1.C.context-d1.tau-source.ci90.0\endcsname{0.062}
\expandafter\def\csname cnt@rev.desc.d1.sspro.p1.C.context-d1.tau-source.ci90.0\endcsname{0}
\expandafter\def\csname pct@rev.desc.d1.sspro.p1.C.context-d1.tau-source.ci90.0\endcsname{6.2}
\expandafter\def\csname ppt@rev.desc.d1.sspro.p1.C.context-d1.tau-source.ci90.0\endcsname{$+$6.2}
\expandafter\def\csname val@rev.desc.d1.sspro.p1.C.context-d1.tau-source.ci90.1\endcsname{0.292}
\expandafter\def\csname cnt@rev.desc.d1.sspro.p1.C.context-d1.tau-source.ci90.1\endcsname{0}
\expandafter\def\csname pct@rev.desc.d1.sspro.p1.C.context-d1.tau-source.ci90.1\endcsname{29.2}
\expandafter\def\csname ppt@rev.desc.d1.sspro.p1.C.context-d1.tau-source.ci90.1\endcsname{$+$29.2}
\expandafter\def\csname val@rev.desc.d1.sspro.p1.C.context-d1.tau-source.ci95-excludes-0\endcsname{1}
\expandafter\def\csname cnt@rev.desc.d1.sspro.p1.C.context-d1.tau-source.ci95-excludes-0\endcsname{1}
\expandafter\def\csname pct@rev.desc.d1.sspro.p1.C.context-d1.tau-source.ci95-excludes-0\endcsname{100.0}
\expandafter\def\csname ppt@rev.desc.d1.sspro.p1.C.context-d1.tau-source.ci95-excludes-0\endcsname{$+$100.0}
\expandafter\def\csname val@rev.desc.d1.sspro.p1.C.context-d1.tau-source.ci95.0\endcsname{0.042}
\expandafter\def\csname cnt@rev.desc.d1.sspro.p1.C.context-d1.tau-source.ci95.0\endcsname{0}
\expandafter\def\csname pct@rev.desc.d1.sspro.p1.C.context-d1.tau-source.ci95.0\endcsname{4.2}
\expandafter\def\csname ppt@rev.desc.d1.sspro.p1.C.context-d1.tau-source.ci95.0\endcsname{$+$4.2}
\expandafter\def\csname val@rev.desc.d1.sspro.p1.C.context-d1.tau-source.ci95.1\endcsname{0.312}
\expandafter\def\csname cnt@rev.desc.d1.sspro.p1.C.context-d1.tau-source.ci95.1\endcsname{0}
\expandafter\def\csname pct@rev.desc.d1.sspro.p1.C.context-d1.tau-source.ci95.1\endcsname{31.2}
\expandafter\def\csname ppt@rev.desc.d1.sspro.p1.C.context-d1.tau-source.ci95.1\endcsname{$+$31.2}
\expandafter\def\csname val@rev.desc.d1.sspro.p1.C.context-d1.tau-source.mean\endcsname{0.167}
\expandafter\def\csname cnt@rev.desc.d1.sspro.p1.C.context-d1.tau-source.mean\endcsname{0}
\expandafter\def\csname pct@rev.desc.d1.sspro.p1.C.context-d1.tau-source.mean\endcsname{16.7}
\expandafter\def\csname ppt@rev.desc.d1.sspro.p1.C.context-d1.tau-source.mean\endcsname{$+$16.7}
\expandafter\def\csname val@rev.desc.d1.sspro.p1.C.context-d1.tau-source.n-apps\endcsname{6}
\expandafter\def\csname cnt@rev.desc.d1.sspro.p1.C.context-d1.tau-source.n-apps\endcsname{6}
\expandafter\def\csname pct@rev.desc.d1.sspro.p1.C.context-d1.tau-source.n-apps\endcsname{600.0}
\expandafter\def\csname ppt@rev.desc.d1.sspro.p1.C.context-d1.tau-source.n-apps\endcsname{$+$600.0}
\expandafter\def\csname val@rev.desc.d1.sspro.p1.C.context-d1.tau-source.tost1.ci90-inside\endcsname{0}
\expandafter\def\csname cnt@rev.desc.d1.sspro.p1.C.context-d1.tau-source.tost1.ci90-inside\endcsname{0}
\expandafter\def\csname pct@rev.desc.d1.sspro.p1.C.context-d1.tau-source.tost1.ci90-inside\endcsname{0.0}
\expandafter\def\csname ppt@rev.desc.d1.sspro.p1.C.context-d1.tau-source.tost1.ci90-inside\endcsname{0.0}
\expandafter\def\csname val@rev.desc.d1.sspro.p1.C.context-d1.tau-source.tost1.equivalent\endcsname{0}
\expandafter\def\csname cnt@rev.desc.d1.sspro.p1.C.context-d1.tau-source.tost1.equivalent\endcsname{0}
\expandafter\def\csname pct@rev.desc.d1.sspro.p1.C.context-d1.tau-source.tost1.equivalent\endcsname{0.0}
\expandafter\def\csname ppt@rev.desc.d1.sspro.p1.C.context-d1.tau-source.tost1.equivalent\endcsname{0.0}
\expandafter\def\csname val@rev.desc.d1.sspro.p1.C.context-d1.tau-source.tost1.p\endcsname{0.997}
\expandafter\def\csname cnt@rev.desc.d1.sspro.p1.C.context-d1.tau-source.tost1.p\endcsname{1}
\expandafter\def\csname pct@rev.desc.d1.sspro.p1.C.context-d1.tau-source.tost1.p\endcsname{99.7}
\expandafter\def\csname ppt@rev.desc.d1.sspro.p1.C.context-d1.tau-source.tost1.p\endcsname{$+$99.7}
\expandafter\def\csname val@rev.desc.d1.sspro.p1.C.context-d1.tau-source.tost5.ci90-inside\endcsname{0}
\expandafter\def\csname cnt@rev.desc.d1.sspro.p1.C.context-d1.tau-source.tost5.ci90-inside\endcsname{0}
\expandafter\def\csname pct@rev.desc.d1.sspro.p1.C.context-d1.tau-source.tost5.ci90-inside\endcsname{0.0}
\expandafter\def\csname ppt@rev.desc.d1.sspro.p1.C.context-d1.tau-source.tost5.ci90-inside\endcsname{0.0}
\expandafter\def\csname val@rev.desc.d1.sspro.p1.C.context-d1.tau-source.tost5.equivalent\endcsname{0}
\expandafter\def\csname cnt@rev.desc.d1.sspro.p1.C.context-d1.tau-source.tost5.equivalent\endcsname{0}
\expandafter\def\csname pct@rev.desc.d1.sspro.p1.C.context-d1.tau-source.tost5.equivalent\endcsname{0.0}
\expandafter\def\csname ppt@rev.desc.d1.sspro.p1.C.context-d1.tau-source.tost5.equivalent\endcsname{0.0}
\expandafter\def\csname val@rev.desc.d1.sspro.p1.C.context-d1.tau-source.tost5.p\endcsname{0.968}
\expandafter\def\csname cnt@rev.desc.d1.sspro.p1.C.context-d1.tau-source.tost5.p\endcsname{1}
\expandafter\def\csname pct@rev.desc.d1.sspro.p1.C.context-d1.tau-source.tost5.p\endcsname{96.8}
\expandafter\def\csname ppt@rev.desc.d1.sspro.p1.C.context-d1.tau-source.tost5.p\endcsname{$+$96.8}
\expandafter\def\csname val@rev.desc.d1.sspro.p1.C.context-d1.tau-target.ci90.0\endcsname{0.042}
\expandafter\def\csname cnt@rev.desc.d1.sspro.p1.C.context-d1.tau-target.ci90.0\endcsname{0}
\expandafter\def\csname pct@rev.desc.d1.sspro.p1.C.context-d1.tau-target.ci90.0\endcsname{4.2}
\expandafter\def\csname ppt@rev.desc.d1.sspro.p1.C.context-d1.tau-target.ci90.0\endcsname{$+$4.2}
\expandafter\def\csname val@rev.desc.d1.sspro.p1.C.context-d1.tau-target.ci90.1\endcsname{0.167}
\expandafter\def\csname cnt@rev.desc.d1.sspro.p1.C.context-d1.tau-target.ci90.1\endcsname{0}
\expandafter\def\csname pct@rev.desc.d1.sspro.p1.C.context-d1.tau-target.ci90.1\endcsname{16.7}
\expandafter\def\csname ppt@rev.desc.d1.sspro.p1.C.context-d1.tau-target.ci90.1\endcsname{$+$16.7}
\expandafter\def\csname val@rev.desc.d1.sspro.p1.C.context-d1.tau-target.ci95-excludes-0\endcsname{1}
\expandafter\def\csname cnt@rev.desc.d1.sspro.p1.C.context-d1.tau-target.ci95-excludes-0\endcsname{1}
\expandafter\def\csname pct@rev.desc.d1.sspro.p1.C.context-d1.tau-target.ci95-excludes-0\endcsname{100.0}
\expandafter\def\csname ppt@rev.desc.d1.sspro.p1.C.context-d1.tau-target.ci95-excludes-0\endcsname{$+$100.0}
\expandafter\def\csname val@rev.desc.d1.sspro.p1.C.context-d1.tau-target.ci95.0\endcsname{0.042}
\expandafter\def\csname cnt@rev.desc.d1.sspro.p1.C.context-d1.tau-target.ci95.0\endcsname{0}
\expandafter\def\csname pct@rev.desc.d1.sspro.p1.C.context-d1.tau-target.ci95.0\endcsname{4.2}
\expandafter\def\csname ppt@rev.desc.d1.sspro.p1.C.context-d1.tau-target.ci95.0\endcsname{$+$4.2}
\expandafter\def\csname val@rev.desc.d1.sspro.p1.C.context-d1.tau-target.ci95.1\endcsname{0.167}
\expandafter\def\csname cnt@rev.desc.d1.sspro.p1.C.context-d1.tau-target.ci95.1\endcsname{0}
\expandafter\def\csname pct@rev.desc.d1.sspro.p1.C.context-d1.tau-target.ci95.1\endcsname{16.7}
\expandafter\def\csname ppt@rev.desc.d1.sspro.p1.C.context-d1.tau-target.ci95.1\endcsname{$+$16.7}
\expandafter\def\csname val@rev.desc.d1.sspro.p1.C.context-d1.tau-target.mean\endcsname{0.104}
\expandafter\def\csname cnt@rev.desc.d1.sspro.p1.C.context-d1.tau-target.mean\endcsname{0}
\expandafter\def\csname pct@rev.desc.d1.sspro.p1.C.context-d1.tau-target.mean\endcsname{10.4}
\expandafter\def\csname ppt@rev.desc.d1.sspro.p1.C.context-d1.tau-target.mean\endcsname{$+$10.4}
\expandafter\def\csname val@rev.desc.d1.sspro.p1.C.context-d1.tau-target.n-apps\endcsname{6}
\expandafter\def\csname cnt@rev.desc.d1.sspro.p1.C.context-d1.tau-target.n-apps\endcsname{6}
\expandafter\def\csname pct@rev.desc.d1.sspro.p1.C.context-d1.tau-target.n-apps\endcsname{600.0}
\expandafter\def\csname ppt@rev.desc.d1.sspro.p1.C.context-d1.tau-target.n-apps\endcsname{$+$600.0}
\expandafter\def\csname val@rev.desc.d1.sspro.p1.C.context-d1.tau-target.tost1.ci90-inside\endcsname{0}
\expandafter\def\csname cnt@rev.desc.d1.sspro.p1.C.context-d1.tau-target.tost1.ci90-inside\endcsname{0}
\expandafter\def\csname pct@rev.desc.d1.sspro.p1.C.context-d1.tau-target.tost1.ci90-inside\endcsname{0.0}
\expandafter\def\csname ppt@rev.desc.d1.sspro.p1.C.context-d1.tau-target.tost1.ci90-inside\endcsname{0.0}
\expandafter\def\csname val@rev.desc.d1.sspro.p1.C.context-d1.tau-target.tost1.equivalent\endcsname{0}
\expandafter\def\csname cnt@rev.desc.d1.sspro.p1.C.context-d1.tau-target.tost1.equivalent\endcsname{0}
\expandafter\def\csname pct@rev.desc.d1.sspro.p1.C.context-d1.tau-target.tost1.equivalent\endcsname{0.0}
\expandafter\def\csname ppt@rev.desc.d1.sspro.p1.C.context-d1.tau-target.tost1.equivalent\endcsname{0.0}
\expandafter\def\csname val@rev.desc.d1.sspro.p1.C.context-d1.tau-target.tost1.p\endcsname{0.999}
\expandafter\def\csname cnt@rev.desc.d1.sspro.p1.C.context-d1.tau-target.tost1.p\endcsname{1}
\expandafter\def\csname pct@rev.desc.d1.sspro.p1.C.context-d1.tau-target.tost1.p\endcsname{99.9}
\expandafter\def\csname ppt@rev.desc.d1.sspro.p1.C.context-d1.tau-target.tost1.p\endcsname{$+$99.9}
\expandafter\def\csname val@rev.desc.d1.sspro.p1.C.context-d1.tau-target.tost5.ci90-inside\endcsname{0}
\expandafter\def\csname cnt@rev.desc.d1.sspro.p1.C.context-d1.tau-target.tost5.ci90-inside\endcsname{0}
\expandafter\def\csname pct@rev.desc.d1.sspro.p1.C.context-d1.tau-target.tost5.ci90-inside\endcsname{0.0}
\expandafter\def\csname ppt@rev.desc.d1.sspro.p1.C.context-d1.tau-target.tost5.ci90-inside\endcsname{0.0}
\expandafter\def\csname val@rev.desc.d1.sspro.p1.C.context-d1.tau-target.tost5.equivalent\endcsname{0}
\expandafter\def\csname cnt@rev.desc.d1.sspro.p1.C.context-d1.tau-target.tost5.equivalent\endcsname{0}
\expandafter\def\csname pct@rev.desc.d1.sspro.p1.C.context-d1.tau-target.tost5.equivalent\endcsname{0.0}
\expandafter\def\csname ppt@rev.desc.d1.sspro.p1.C.context-d1.tau-target.tost5.equivalent\endcsname{0.0}
\expandafter\def\csname val@rev.desc.d1.sspro.p1.C.context-d1.tau-target.tost5.p\endcsname{0.931}
\expandafter\def\csname cnt@rev.desc.d1.sspro.p1.C.context-d1.tau-target.tost5.p\endcsname{1}
\expandafter\def\csname pct@rev.desc.d1.sspro.p1.C.context-d1.tau-target.tost5.p\endcsname{93.1}
\expandafter\def\csname ppt@rev.desc.d1.sspro.p1.C.context-d1.tau-target.tost5.p\endcsname{$+$93.1}
\expandafter\def\csname val@rev.desc.d1.sspro.p1.C.context-d1.zero-estimate\endcsname{0}
\expandafter\def\csname cnt@rev.desc.d1.sspro.p1.C.context-d1.zero-estimate\endcsname{0}
\expandafter\def\csname pct@rev.desc.d1.sspro.p1.C.context-d1.zero-estimate\endcsname{0.0}
\expandafter\def\csname ppt@rev.desc.d1.sspro.p1.C.context-d1.zero-estimate\endcsname{0.0}
\expandafter\def\csname val@rev.desc.d1.sspro.p1.C.dual-d2.delta-tau.ci90.0\endcsname{-0.167}
\expandafter\def\csname cnt@rev.desc.d1.sspro.p1.C.dual-d2.delta-tau.ci90.0\endcsname{0}
\expandafter\def\csname pct@rev.desc.d1.sspro.p1.C.dual-d2.delta-tau.ci90.0\endcsname{$-$16.7}
\expandafter\def\csname ppt@rev.desc.d1.sspro.p1.C.dual-d2.delta-tau.ci90.0\endcsname{$-$16.7}
\expandafter\def\csname val@rev.desc.d1.sspro.p1.C.dual-d2.delta-tau.ci90.1\endcsname{0.167}
\expandafter\def\csname cnt@rev.desc.d1.sspro.p1.C.dual-d2.delta-tau.ci90.1\endcsname{0}
\expandafter\def\csname pct@rev.desc.d1.sspro.p1.C.dual-d2.delta-tau.ci90.1\endcsname{16.7}
\expandafter\def\csname ppt@rev.desc.d1.sspro.p1.C.dual-d2.delta-tau.ci90.1\endcsname{$+$16.7}
\expandafter\def\csname val@rev.desc.d1.sspro.p1.C.dual-d2.delta-tau.ci95-excludes-0\endcsname{0}
\expandafter\def\csname cnt@rev.desc.d1.sspro.p1.C.dual-d2.delta-tau.ci95-excludes-0\endcsname{0}
\expandafter\def\csname pct@rev.desc.d1.sspro.p1.C.dual-d2.delta-tau.ci95-excludes-0\endcsname{0.0}
\expandafter\def\csname ppt@rev.desc.d1.sspro.p1.C.dual-d2.delta-tau.ci95-excludes-0\endcsname{0.0}
\expandafter\def\csname val@rev.desc.d1.sspro.p1.C.dual-d2.delta-tau.ci95.0\endcsname{-0.229}
\expandafter\def\csname cnt@rev.desc.d1.sspro.p1.C.dual-d2.delta-tau.ci95.0\endcsname{0}
\expandafter\def\csname pct@rev.desc.d1.sspro.p1.C.dual-d2.delta-tau.ci95.0\endcsname{$-$22.9}
\expandafter\def\csname ppt@rev.desc.d1.sspro.p1.C.dual-d2.delta-tau.ci95.0\endcsname{$-$22.9}
\expandafter\def\csname val@rev.desc.d1.sspro.p1.C.dual-d2.delta-tau.ci95.1\endcsname{0.188}
\expandafter\def\csname cnt@rev.desc.d1.sspro.p1.C.dual-d2.delta-tau.ci95.1\endcsname{0}
\expandafter\def\csname pct@rev.desc.d1.sspro.p1.C.dual-d2.delta-tau.ci95.1\endcsname{18.8}
\expandafter\def\csname ppt@rev.desc.d1.sspro.p1.C.dual-d2.delta-tau.ci95.1\endcsname{$+$18.8}
\expandafter\def\csname val@rev.desc.d1.sspro.p1.C.dual-d2.delta-tau.mean\endcsname{0.000}
\expandafter\def\csname cnt@rev.desc.d1.sspro.p1.C.dual-d2.delta-tau.mean\endcsname{0}
\expandafter\def\csname pct@rev.desc.d1.sspro.p1.C.dual-d2.delta-tau.mean\endcsname{0.0}
\expandafter\def\csname ppt@rev.desc.d1.sspro.p1.C.dual-d2.delta-tau.mean\endcsname{0.0}
\expandafter\def\csname val@rev.desc.d1.sspro.p1.C.dual-d2.delta-tau.n-apps\endcsname{6}
\expandafter\def\csname cnt@rev.desc.d1.sspro.p1.C.dual-d2.delta-tau.n-apps\endcsname{6}
\expandafter\def\csname pct@rev.desc.d1.sspro.p1.C.dual-d2.delta-tau.n-apps\endcsname{600.0}
\expandafter\def\csname ppt@rev.desc.d1.sspro.p1.C.dual-d2.delta-tau.n-apps\endcsname{$+$600.0}
\expandafter\def\csname val@rev.desc.d1.sspro.p1.C.dual-d2.delta-tau.tost1.ci90-inside\endcsname{0}
\expandafter\def\csname cnt@rev.desc.d1.sspro.p1.C.dual-d2.delta-tau.tost1.ci90-inside\endcsname{0}
\expandafter\def\csname pct@rev.desc.d1.sspro.p1.C.dual-d2.delta-tau.tost1.ci90-inside\endcsname{0.0}
\expandafter\def\csname ppt@rev.desc.d1.sspro.p1.C.dual-d2.delta-tau.tost1.ci90-inside\endcsname{0.0}
\expandafter\def\csname val@rev.desc.d1.sspro.p1.C.dual-d2.delta-tau.tost1.equivalent\endcsname{0}
\expandafter\def\csname cnt@rev.desc.d1.sspro.p1.C.dual-d2.delta-tau.tost1.equivalent\endcsname{0}
\expandafter\def\csname pct@rev.desc.d1.sspro.p1.C.dual-d2.delta-tau.tost1.equivalent\endcsname{0.0}
\expandafter\def\csname ppt@rev.desc.d1.sspro.p1.C.dual-d2.delta-tau.tost1.equivalent\endcsname{0.0}
\expandafter\def\csname val@rev.desc.d1.sspro.p1.C.dual-d2.delta-tau.tost1.p\endcsname{0.476}
\expandafter\def\csname cnt@rev.desc.d1.sspro.p1.C.dual-d2.delta-tau.tost1.p\endcsname{0}
\expandafter\def\csname pct@rev.desc.d1.sspro.p1.C.dual-d2.delta-tau.tost1.p\endcsname{47.6}
\expandafter\def\csname ppt@rev.desc.d1.sspro.p1.C.dual-d2.delta-tau.tost1.p\endcsname{$+$47.6}
\expandafter\def\csname val@rev.desc.d1.sspro.p1.C.dual-d2.delta-tau.tost5.ci90-inside\endcsname{0}
\expandafter\def\csname cnt@rev.desc.d1.sspro.p1.C.dual-d2.delta-tau.tost5.ci90-inside\endcsname{0}
\expandafter\def\csname pct@rev.desc.d1.sspro.p1.C.dual-d2.delta-tau.tost5.ci90-inside\endcsname{0.0}
\expandafter\def\csname ppt@rev.desc.d1.sspro.p1.C.dual-d2.delta-tau.tost5.ci90-inside\endcsname{0.0}
\expandafter\def\csname val@rev.desc.d1.sspro.p1.C.dual-d2.delta-tau.tost5.equivalent\endcsname{0}
\expandafter\def\csname cnt@rev.desc.d1.sspro.p1.C.dual-d2.delta-tau.tost5.equivalent\endcsname{0}
\expandafter\def\csname pct@rev.desc.d1.sspro.p1.C.dual-d2.delta-tau.tost5.equivalent\endcsname{0.0}
\expandafter\def\csname ppt@rev.desc.d1.sspro.p1.C.dual-d2.delta-tau.tost5.equivalent\endcsname{0.0}
\expandafter\def\csname val@rev.desc.d1.sspro.p1.C.dual-d2.delta-tau.tost5.p\endcsname{0.329}
\expandafter\def\csname cnt@rev.desc.d1.sspro.p1.C.dual-d2.delta-tau.tost5.p\endcsname{0}
\expandafter\def\csname pct@rev.desc.d1.sspro.p1.C.dual-d2.delta-tau.tost5.p\endcsname{32.9}
\expandafter\def\csname ppt@rev.desc.d1.sspro.p1.C.dual-d2.delta-tau.tost5.p\endcsname{$+$32.9}
\expandafter\def\csname val@rev.desc.d1.sspro.p1.C.dual-d2.material-flip\endcsname{0}
\expandafter\def\csname cnt@rev.desc.d1.sspro.p1.C.dual-d2.material-flip\endcsname{0}
\expandafter\def\csname pct@rev.desc.d1.sspro.p1.C.dual-d2.material-flip\endcsname{0.0}
\expandafter\def\csname ppt@rev.desc.d1.sspro.p1.C.dual-d2.material-flip\endcsname{0.0}
\expandafter\def\csname val@rev.desc.d1.sspro.p1.C.dual-d2.n-apps\endcsname{6}
\expandafter\def\csname cnt@rev.desc.d1.sspro.p1.C.dual-d2.n-apps\endcsname{6}
\expandafter\def\csname pct@rev.desc.d1.sspro.p1.C.dual-d2.n-apps\endcsname{600.0}
\expandafter\def\csname ppt@rev.desc.d1.sspro.p1.C.dual-d2.n-apps\endcsname{$+$600.0}
\expandafter\def\csname val@rev.desc.d1.sspro.p1.C.dual-d2.n-tasks\endcsname{48}
\expandafter\def\csname cnt@rev.desc.d1.sspro.p1.C.dual-d2.n-tasks\endcsname{48}
\expandafter\def\csname pct@rev.desc.d1.sspro.p1.C.dual-d2.n-tasks\endcsname{4800.0}
\expandafter\def\csname ppt@rev.desc.d1.sspro.p1.C.dual-d2.n-tasks\endcsname{$+$4800.0}
\expandafter\def\csname val@rev.desc.d1.sspro.p1.C.dual-d2.sign-differs\endcsname{0}
\expandafter\def\csname cnt@rev.desc.d1.sspro.p1.C.dual-d2.sign-differs\endcsname{0}
\expandafter\def\csname pct@rev.desc.d1.sspro.p1.C.dual-d2.sign-differs\endcsname{0.0}
\expandafter\def\csname ppt@rev.desc.d1.sspro.p1.C.dual-d2.sign-differs\endcsname{0.0}
\expandafter\def\csname val@rev.desc.d1.sspro.p1.C.dual-d2.tau-source.ci90.0\endcsname{-0.021}
\expandafter\def\csname cnt@rev.desc.d1.sspro.p1.C.dual-d2.tau-source.ci90.0\endcsname{0}
\expandafter\def\csname pct@rev.desc.d1.sspro.p1.C.dual-d2.tau-source.ci90.0\endcsname{$-$2.1}
\expandafter\def\csname ppt@rev.desc.d1.sspro.p1.C.dual-d2.tau-source.ci90.0\endcsname{$-$2.1}
\expandafter\def\csname val@rev.desc.d1.sspro.p1.C.dual-d2.tau-source.ci90.1\endcsname{0.271}
\expandafter\def\csname cnt@rev.desc.d1.sspro.p1.C.dual-d2.tau-source.ci90.1\endcsname{0}
\expandafter\def\csname pct@rev.desc.d1.sspro.p1.C.dual-d2.tau-source.ci90.1\endcsname{27.1}
\expandafter\def\csname ppt@rev.desc.d1.sspro.p1.C.dual-d2.tau-source.ci90.1\endcsname{$+$27.1}
\expandafter\def\csname val@rev.desc.d1.sspro.p1.C.dual-d2.tau-source.ci95-excludes-0\endcsname{0}
\expandafter\def\csname cnt@rev.desc.d1.sspro.p1.C.dual-d2.tau-source.ci95-excludes-0\endcsname{0}
\expandafter\def\csname pct@rev.desc.d1.sspro.p1.C.dual-d2.tau-source.ci95-excludes-0\endcsname{0.0}
\expandafter\def\csname ppt@rev.desc.d1.sspro.p1.C.dual-d2.tau-source.ci95-excludes-0\endcsname{0.0}
\expandafter\def\csname val@rev.desc.d1.sspro.p1.C.dual-d2.tau-source.ci95.0\endcsname{-0.042}
\expandafter\def\csname cnt@rev.desc.d1.sspro.p1.C.dual-d2.tau-source.ci95.0\endcsname{0}
\expandafter\def\csname pct@rev.desc.d1.sspro.p1.C.dual-d2.tau-source.ci95.0\endcsname{$-$4.2}
\expandafter\def\csname ppt@rev.desc.d1.sspro.p1.C.dual-d2.tau-source.ci95.0\endcsname{$-$4.2}
\expandafter\def\csname val@rev.desc.d1.sspro.p1.C.dual-d2.tau-source.ci95.1\endcsname{0.292}
\expandafter\def\csname cnt@rev.desc.d1.sspro.p1.C.dual-d2.tau-source.ci95.1\endcsname{0}
\expandafter\def\csname pct@rev.desc.d1.sspro.p1.C.dual-d2.tau-source.ci95.1\endcsname{29.2}
\expandafter\def\csname ppt@rev.desc.d1.sspro.p1.C.dual-d2.tau-source.ci95.1\endcsname{$+$29.2}
\expandafter\def\csname val@rev.desc.d1.sspro.p1.C.dual-d2.tau-source.mean\endcsname{0.125}
\expandafter\def\csname cnt@rev.desc.d1.sspro.p1.C.dual-d2.tau-source.mean\endcsname{0}
\expandafter\def\csname pct@rev.desc.d1.sspro.p1.C.dual-d2.tau-source.mean\endcsname{12.5}
\expandafter\def\csname ppt@rev.desc.d1.sspro.p1.C.dual-d2.tau-source.mean\endcsname{$+$12.5}
\expandafter\def\csname val@rev.desc.d1.sspro.p1.C.dual-d2.tau-source.n-apps\endcsname{6}
\expandafter\def\csname cnt@rev.desc.d1.sspro.p1.C.dual-d2.tau-source.n-apps\endcsname{6}
\expandafter\def\csname pct@rev.desc.d1.sspro.p1.C.dual-d2.tau-source.n-apps\endcsname{600.0}
\expandafter\def\csname ppt@rev.desc.d1.sspro.p1.C.dual-d2.tau-source.n-apps\endcsname{$+$600.0}
\expandafter\def\csname val@rev.desc.d1.sspro.p1.C.dual-d2.tau-source.tost1.ci90-inside\endcsname{0}
\expandafter\def\csname cnt@rev.desc.d1.sspro.p1.C.dual-d2.tau-source.tost1.ci90-inside\endcsname{0}
\expandafter\def\csname pct@rev.desc.d1.sspro.p1.C.dual-d2.tau-source.tost1.ci90-inside\endcsname{0.0}
\expandafter\def\csname ppt@rev.desc.d1.sspro.p1.C.dual-d2.tau-source.tost1.ci90-inside\endcsname{0.0}
\expandafter\def\csname val@rev.desc.d1.sspro.p1.C.dual-d2.tau-source.tost1.equivalent\endcsname{0}
\expandafter\def\csname cnt@rev.desc.d1.sspro.p1.C.dual-d2.tau-source.tost1.equivalent\endcsname{0}
\expandafter\def\csname pct@rev.desc.d1.sspro.p1.C.dual-d2.tau-source.tost1.equivalent\endcsname{0.0}
\expandafter\def\csname ppt@rev.desc.d1.sspro.p1.C.dual-d2.tau-source.tost1.equivalent\endcsname{0.0}
\expandafter\def\csname val@rev.desc.d1.sspro.p1.C.dual-d2.tau-source.tost1.p\endcsname{0.904}
\expandafter\def\csname cnt@rev.desc.d1.sspro.p1.C.dual-d2.tau-source.tost1.p\endcsname{1}
\expandafter\def\csname pct@rev.desc.d1.sspro.p1.C.dual-d2.tau-source.tost1.p\endcsname{90.4}
\expandafter\def\csname ppt@rev.desc.d1.sspro.p1.C.dual-d2.tau-source.tost1.p\endcsname{$+$90.4}
\expandafter\def\csname val@rev.desc.d1.sspro.p1.C.dual-d2.tau-source.tost5.ci90-inside\endcsname{0}
\expandafter\def\csname cnt@rev.desc.d1.sspro.p1.C.dual-d2.tau-source.tost5.ci90-inside\endcsname{0}
\expandafter\def\csname pct@rev.desc.d1.sspro.p1.C.dual-d2.tau-source.tost5.ci90-inside\endcsname{0.0}
\expandafter\def\csname ppt@rev.desc.d1.sspro.p1.C.dual-d2.tau-source.tost5.ci90-inside\endcsname{0.0}
\expandafter\def\csname val@rev.desc.d1.sspro.p1.C.dual-d2.tau-source.tost5.equivalent\endcsname{0}
\expandafter\def\csname cnt@rev.desc.d1.sspro.p1.C.dual-d2.tau-source.tost5.equivalent\endcsname{0}
\expandafter\def\csname pct@rev.desc.d1.sspro.p1.C.dual-d2.tau-source.tost5.equivalent\endcsname{0.0}
\expandafter\def\csname ppt@rev.desc.d1.sspro.p1.C.dual-d2.tau-source.tost5.equivalent\endcsname{0.0}
\expandafter\def\csname val@rev.desc.d1.sspro.p1.C.dual-d2.tau-source.tost5.p\endcsname{0.806}
\expandafter\def\csname cnt@rev.desc.d1.sspro.p1.C.dual-d2.tau-source.tost5.p\endcsname{1}
\expandafter\def\csname pct@rev.desc.d1.sspro.p1.C.dual-d2.tau-source.tost5.p\endcsname{80.6}
\expandafter\def\csname ppt@rev.desc.d1.sspro.p1.C.dual-d2.tau-source.tost5.p\endcsname{$+$80.6}
\expandafter\def\csname val@rev.desc.d1.sspro.p1.C.dual-d2.tau-target.ci90.0\endcsname{0.021}
\expandafter\def\csname cnt@rev.desc.d1.sspro.p1.C.dual-d2.tau-target.ci90.0\endcsname{0}
\expandafter\def\csname pct@rev.desc.d1.sspro.p1.C.dual-d2.tau-target.ci90.0\endcsname{2.1}
\expandafter\def\csname ppt@rev.desc.d1.sspro.p1.C.dual-d2.tau-target.ci90.0\endcsname{$+$2.1}
\expandafter\def\csname val@rev.desc.d1.sspro.p1.C.dual-d2.tau-target.ci90.1\endcsname{0.229}
\expandafter\def\csname cnt@rev.desc.d1.sspro.p1.C.dual-d2.tau-target.ci90.1\endcsname{0}
\expandafter\def\csname pct@rev.desc.d1.sspro.p1.C.dual-d2.tau-target.ci90.1\endcsname{22.9}
\expandafter\def\csname ppt@rev.desc.d1.sspro.p1.C.dual-d2.tau-target.ci90.1\endcsname{$+$22.9}
\expandafter\def\csname val@rev.desc.d1.sspro.p1.C.dual-d2.tau-target.ci95-excludes-0\endcsname{0}
\expandafter\def\csname cnt@rev.desc.d1.sspro.p1.C.dual-d2.tau-target.ci95-excludes-0\endcsname{0}
\expandafter\def\csname pct@rev.desc.d1.sspro.p1.C.dual-d2.tau-target.ci95-excludes-0\endcsname{0.0}
\expandafter\def\csname ppt@rev.desc.d1.sspro.p1.C.dual-d2.tau-target.ci95-excludes-0\endcsname{0.0}
\expandafter\def\csname val@rev.desc.d1.sspro.p1.C.dual-d2.tau-target.ci95.0\endcsname{0.000}
\expandafter\def\csname cnt@rev.desc.d1.sspro.p1.C.dual-d2.tau-target.ci95.0\endcsname{0}
\expandafter\def\csname pct@rev.desc.d1.sspro.p1.C.dual-d2.tau-target.ci95.0\endcsname{0.0}
\expandafter\def\csname ppt@rev.desc.d1.sspro.p1.C.dual-d2.tau-target.ci95.0\endcsname{0.0}
\expandafter\def\csname val@rev.desc.d1.sspro.p1.C.dual-d2.tau-target.ci95.1\endcsname{0.250}
\expandafter\def\csname cnt@rev.desc.d1.sspro.p1.C.dual-d2.tau-target.ci95.1\endcsname{0}
\expandafter\def\csname pct@rev.desc.d1.sspro.p1.C.dual-d2.tau-target.ci95.1\endcsname{25.0}
\expandafter\def\csname ppt@rev.desc.d1.sspro.p1.C.dual-d2.tau-target.ci95.1\endcsname{$+$25.0}
\expandafter\def\csname val@rev.desc.d1.sspro.p1.C.dual-d2.tau-target.mean\endcsname{0.125}
\expandafter\def\csname cnt@rev.desc.d1.sspro.p1.C.dual-d2.tau-target.mean\endcsname{0}
\expandafter\def\csname pct@rev.desc.d1.sspro.p1.C.dual-d2.tau-target.mean\endcsname{12.5}
\expandafter\def\csname ppt@rev.desc.d1.sspro.p1.C.dual-d2.tau-target.mean\endcsname{$+$12.5}
\expandafter\def\csname val@rev.desc.d1.sspro.p1.C.dual-d2.tau-target.n-apps\endcsname{6}
\expandafter\def\csname cnt@rev.desc.d1.sspro.p1.C.dual-d2.tau-target.n-apps\endcsname{6}
\expandafter\def\csname pct@rev.desc.d1.sspro.p1.C.dual-d2.tau-target.n-apps\endcsname{600.0}
\expandafter\def\csname ppt@rev.desc.d1.sspro.p1.C.dual-d2.tau-target.n-apps\endcsname{$+$600.0}
\expandafter\def\csname val@rev.desc.d1.sspro.p1.C.dual-d2.tau-target.tost1.ci90-inside\endcsname{0}
\expandafter\def\csname cnt@rev.desc.d1.sspro.p1.C.dual-d2.tau-target.tost1.ci90-inside\endcsname{0}
\expandafter\def\csname pct@rev.desc.d1.sspro.p1.C.dual-d2.tau-target.tost1.ci90-inside\endcsname{0.0}
\expandafter\def\csname ppt@rev.desc.d1.sspro.p1.C.dual-d2.tau-target.tost1.ci90-inside\endcsname{0.0}
\expandafter\def\csname val@rev.desc.d1.sspro.p1.C.dual-d2.tau-target.tost1.equivalent\endcsname{0}
\expandafter\def\csname cnt@rev.desc.d1.sspro.p1.C.dual-d2.tau-target.tost1.equivalent\endcsname{0}
\expandafter\def\csname pct@rev.desc.d1.sspro.p1.C.dual-d2.tau-target.tost1.equivalent\endcsname{0.0}
\expandafter\def\csname ppt@rev.desc.d1.sspro.p1.C.dual-d2.tau-target.tost1.equivalent\endcsname{0.0}
\expandafter\def\csname val@rev.desc.d1.sspro.p1.C.dual-d2.tau-target.tost1.p\endcsname{0.958}
\expandafter\def\csname cnt@rev.desc.d1.sspro.p1.C.dual-d2.tau-target.tost1.p\endcsname{1}
\expandafter\def\csname pct@rev.desc.d1.sspro.p1.C.dual-d2.tau-target.tost1.p\endcsname{95.8}
\expandafter\def\csname ppt@rev.desc.d1.sspro.p1.C.dual-d2.tau-target.tost1.p\endcsname{$+$95.8}
\expandafter\def\csname val@rev.desc.d1.sspro.p1.C.dual-d2.tau-target.tost5.ci90-inside\endcsname{0}
\expandafter\def\csname cnt@rev.desc.d1.sspro.p1.C.dual-d2.tau-target.tost5.ci90-inside\endcsname{0}
\expandafter\def\csname pct@rev.desc.d1.sspro.p1.C.dual-d2.tau-target.tost5.ci90-inside\endcsname{0.0}
\expandafter\def\csname ppt@rev.desc.d1.sspro.p1.C.dual-d2.tau-target.tost5.ci90-inside\endcsname{0.0}
\expandafter\def\csname val@rev.desc.d1.sspro.p1.C.dual-d2.tau-target.tost5.equivalent\endcsname{0}
\expandafter\def\csname cnt@rev.desc.d1.sspro.p1.C.dual-d2.tau-target.tost5.equivalent\endcsname{0}
\expandafter\def\csname pct@rev.desc.d1.sspro.p1.C.dual-d2.tau-target.tost5.equivalent\endcsname{0.0}
\expandafter\def\csname ppt@rev.desc.d1.sspro.p1.C.dual-d2.tau-target.tost5.equivalent\endcsname{0.0}
\expandafter\def\csname val@rev.desc.d1.sspro.p1.C.dual-d2.tau-target.tost5.p\endcsname{0.857}
\expandafter\def\csname cnt@rev.desc.d1.sspro.p1.C.dual-d2.tau-target.tost5.p\endcsname{1}
\expandafter\def\csname pct@rev.desc.d1.sspro.p1.C.dual-d2.tau-target.tost5.p\endcsname{85.7}
\expandafter\def\csname ppt@rev.desc.d1.sspro.p1.C.dual-d2.tau-target.tost5.p\endcsname{$+$85.7}
\expandafter\def\csname val@rev.desc.d1.sspro.p1.C.dual-d2.zero-estimate\endcsname{0}
\expandafter\def\csname cnt@rev.desc.d1.sspro.p1.C.dual-d2.zero-estimate\endcsname{0}
\expandafter\def\csname pct@rev.desc.d1.sspro.p1.C.dual-d2.zero-estimate\endcsname{0.0}
\expandafter\def\csname ppt@rev.desc.d1.sspro.p1.C.dual-d2.zero-estimate\endcsname{0.0}
\expandafter\def\csname val@rev.desc.d1.sspro.p1.C.grid2-d3.delta-tau.ci90.0\endcsname{-0.104}
\expandafter\def\csname cnt@rev.desc.d1.sspro.p1.C.grid2-d3.delta-tau.ci90.0\endcsname{0}
\expandafter\def\csname pct@rev.desc.d1.sspro.p1.C.grid2-d3.delta-tau.ci90.0\endcsname{$-$10.4}
\expandafter\def\csname ppt@rev.desc.d1.sspro.p1.C.grid2-d3.delta-tau.ci90.0\endcsname{$-$10.4}
\expandafter\def\csname val@rev.desc.d1.sspro.p1.C.grid2-d3.delta-tau.ci90.1\endcsname{-0.021}
\expandafter\def\csname cnt@rev.desc.d1.sspro.p1.C.grid2-d3.delta-tau.ci90.1\endcsname{0}
\expandafter\def\csname pct@rev.desc.d1.sspro.p1.C.grid2-d3.delta-tau.ci90.1\endcsname{$-$2.1}
\expandafter\def\csname ppt@rev.desc.d1.sspro.p1.C.grid2-d3.delta-tau.ci90.1\endcsname{$-$2.1}
\expandafter\def\csname val@rev.desc.d1.sspro.p1.C.grid2-d3.delta-tau.ci95-excludes-0\endcsname{1}
\expandafter\def\csname cnt@rev.desc.d1.sspro.p1.C.grid2-d3.delta-tau.ci95-excludes-0\endcsname{1}
\expandafter\def\csname pct@rev.desc.d1.sspro.p1.C.grid2-d3.delta-tau.ci95-excludes-0\endcsname{100.0}
\expandafter\def\csname ppt@rev.desc.d1.sspro.p1.C.grid2-d3.delta-tau.ci95-excludes-0\endcsname{$+$100.0}
\expandafter\def\csname val@rev.desc.d1.sspro.p1.C.grid2-d3.delta-tau.ci95.0\endcsname{-0.104}
\expandafter\def\csname cnt@rev.desc.d1.sspro.p1.C.grid2-d3.delta-tau.ci95.0\endcsname{0}
\expandafter\def\csname pct@rev.desc.d1.sspro.p1.C.grid2-d3.delta-tau.ci95.0\endcsname{$-$10.4}
\expandafter\def\csname ppt@rev.desc.d1.sspro.p1.C.grid2-d3.delta-tau.ci95.0\endcsname{$-$10.4}
\expandafter\def\csname val@rev.desc.d1.sspro.p1.C.grid2-d3.delta-tau.ci95.1\endcsname{-0.021}
\expandafter\def\csname cnt@rev.desc.d1.sspro.p1.C.grid2-d3.delta-tau.ci95.1\endcsname{0}
\expandafter\def\csname pct@rev.desc.d1.sspro.p1.C.grid2-d3.delta-tau.ci95.1\endcsname{$-$2.1}
\expandafter\def\csname ppt@rev.desc.d1.sspro.p1.C.grid2-d3.delta-tau.ci95.1\endcsname{$-$2.1}
\expandafter\def\csname val@rev.desc.d1.sspro.p1.C.grid2-d3.delta-tau.mean\endcsname{-0.062}
\expandafter\def\csname cnt@rev.desc.d1.sspro.p1.C.grid2-d3.delta-tau.mean\endcsname{0}
\expandafter\def\csname pct@rev.desc.d1.sspro.p1.C.grid2-d3.delta-tau.mean\endcsname{$-$6.2}
\expandafter\def\csname ppt@rev.desc.d1.sspro.p1.C.grid2-d3.delta-tau.mean\endcsname{$-$6.2}
\expandafter\def\csname val@rev.desc.d1.sspro.p1.C.grid2-d3.delta-tau.n-apps\endcsname{6}
\expandafter\def\csname cnt@rev.desc.d1.sspro.p1.C.grid2-d3.delta-tau.n-apps\endcsname{6}
\expandafter\def\csname pct@rev.desc.d1.sspro.p1.C.grid2-d3.delta-tau.n-apps\endcsname{600.0}
\expandafter\def\csname ppt@rev.desc.d1.sspro.p1.C.grid2-d3.delta-tau.n-apps\endcsname{$+$600.0}
\expandafter\def\csname val@rev.desc.d1.sspro.p1.C.grid2-d3.delta-tau.tost1.ci90-inside\endcsname{0}
\expandafter\def\csname cnt@rev.desc.d1.sspro.p1.C.grid2-d3.delta-tau.tost1.ci90-inside\endcsname{0}
\expandafter\def\csname pct@rev.desc.d1.sspro.p1.C.grid2-d3.delta-tau.tost1.ci90-inside\endcsname{0.0}
\expandafter\def\csname ppt@rev.desc.d1.sspro.p1.C.grid2-d3.delta-tau.tost1.ci90-inside\endcsname{0.0}
\expandafter\def\csname val@rev.desc.d1.sspro.p1.C.grid2-d3.delta-tau.tost1.equivalent\endcsname{0}
\expandafter\def\csname cnt@rev.desc.d1.sspro.p1.C.grid2-d3.delta-tau.tost1.equivalent\endcsname{0}
\expandafter\def\csname pct@rev.desc.d1.sspro.p1.C.grid2-d3.delta-tau.tost1.equivalent\endcsname{0.0}
\expandafter\def\csname ppt@rev.desc.d1.sspro.p1.C.grid2-d3.delta-tau.tost1.equivalent\endcsname{0.0}
\expandafter\def\csname val@rev.desc.d1.sspro.p1.C.grid2-d3.delta-tau.tost1.p\endcsname{0.982}
\expandafter\def\csname cnt@rev.desc.d1.sspro.p1.C.grid2-d3.delta-tau.tost1.p\endcsname{1}
\expandafter\def\csname pct@rev.desc.d1.sspro.p1.C.grid2-d3.delta-tau.tost1.p\endcsname{98.2}
\expandafter\def\csname ppt@rev.desc.d1.sspro.p1.C.grid2-d3.delta-tau.tost1.p\endcsname{$+$98.2}
\expandafter\def\csname val@rev.desc.d1.sspro.p1.C.grid2-d3.delta-tau.tost5.ci90-inside\endcsname{0}
\expandafter\def\csname cnt@rev.desc.d1.sspro.p1.C.grid2-d3.delta-tau.tost5.ci90-inside\endcsname{0}
\expandafter\def\csname pct@rev.desc.d1.sspro.p1.C.grid2-d3.delta-tau.tost5.ci90-inside\endcsname{0.0}
\expandafter\def\csname ppt@rev.desc.d1.sspro.p1.C.grid2-d3.delta-tau.tost5.ci90-inside\endcsname{0.0}
\expandafter\def\csname val@rev.desc.d1.sspro.p1.C.grid2-d3.delta-tau.tost5.equivalent\endcsname{0}
\expandafter\def\csname cnt@rev.desc.d1.sspro.p1.C.grid2-d3.delta-tau.tost5.equivalent\endcsname{0}
\expandafter\def\csname pct@rev.desc.d1.sspro.p1.C.grid2-d3.delta-tau.tost5.equivalent\endcsname{0.0}
\expandafter\def\csname ppt@rev.desc.d1.sspro.p1.C.grid2-d3.delta-tau.tost5.equivalent\endcsname{0.0}
\expandafter\def\csname val@rev.desc.d1.sspro.p1.C.grid2-d3.delta-tau.tost5.p\endcsname{0.656}
\expandafter\def\csname cnt@rev.desc.d1.sspro.p1.C.grid2-d3.delta-tau.tost5.p\endcsname{1}
\expandafter\def\csname pct@rev.desc.d1.sspro.p1.C.grid2-d3.delta-tau.tost5.p\endcsname{65.6}
\expandafter\def\csname ppt@rev.desc.d1.sspro.p1.C.grid2-d3.delta-tau.tost5.p\endcsname{$+$65.6}
\expandafter\def\csname val@rev.desc.d1.sspro.p1.C.grid2-d3.material-flip\endcsname{0}
\expandafter\def\csname cnt@rev.desc.d1.sspro.p1.C.grid2-d3.material-flip\endcsname{0}
\expandafter\def\csname pct@rev.desc.d1.sspro.p1.C.grid2-d3.material-flip\endcsname{0.0}
\expandafter\def\csname ppt@rev.desc.d1.sspro.p1.C.grid2-d3.material-flip\endcsname{0.0}
\expandafter\def\csname val@rev.desc.d1.sspro.p1.C.grid2-d3.n-apps\endcsname{6}
\expandafter\def\csname cnt@rev.desc.d1.sspro.p1.C.grid2-d3.n-apps\endcsname{6}
\expandafter\def\csname pct@rev.desc.d1.sspro.p1.C.grid2-d3.n-apps\endcsname{600.0}
\expandafter\def\csname ppt@rev.desc.d1.sspro.p1.C.grid2-d3.n-apps\endcsname{$+$600.0}
\expandafter\def\csname val@rev.desc.d1.sspro.p1.C.grid2-d3.n-tasks\endcsname{48}
\expandafter\def\csname cnt@rev.desc.d1.sspro.p1.C.grid2-d3.n-tasks\endcsname{48}
\expandafter\def\csname pct@rev.desc.d1.sspro.p1.C.grid2-d3.n-tasks\endcsname{4800.0}
\expandafter\def\csname ppt@rev.desc.d1.sspro.p1.C.grid2-d3.n-tasks\endcsname{$+$4800.0}
\expandafter\def\csname val@rev.desc.d1.sspro.p1.C.grid2-d3.sign-differs\endcsname{0}
\expandafter\def\csname cnt@rev.desc.d1.sspro.p1.C.grid2-d3.sign-differs\endcsname{0}
\expandafter\def\csname pct@rev.desc.d1.sspro.p1.C.grid2-d3.sign-differs\endcsname{0.0}
\expandafter\def\csname ppt@rev.desc.d1.sspro.p1.C.grid2-d3.sign-differs\endcsname{0.0}
\expandafter\def\csname val@rev.desc.d1.sspro.p1.C.grid2-d3.tau-source.ci90.0\endcsname{0.083}
\expandafter\def\csname cnt@rev.desc.d1.sspro.p1.C.grid2-d3.tau-source.ci90.0\endcsname{0}
\expandafter\def\csname pct@rev.desc.d1.sspro.p1.C.grid2-d3.tau-source.ci90.0\endcsname{8.3}
\expandafter\def\csname ppt@rev.desc.d1.sspro.p1.C.grid2-d3.tau-source.ci90.0\endcsname{$+$8.3}
\expandafter\def\csname val@rev.desc.d1.sspro.p1.C.grid2-d3.tau-source.ci90.1\endcsname{0.333}
\expandafter\def\csname cnt@rev.desc.d1.sspro.p1.C.grid2-d3.tau-source.ci90.1\endcsname{0}
\expandafter\def\csname pct@rev.desc.d1.sspro.p1.C.grid2-d3.tau-source.ci90.1\endcsname{33.3}
\expandafter\def\csname ppt@rev.desc.d1.sspro.p1.C.grid2-d3.tau-source.ci90.1\endcsname{$+$33.3}
\expandafter\def\csname val@rev.desc.d1.sspro.p1.C.grid2-d3.tau-source.ci95-excludes-0\endcsname{1}
\expandafter\def\csname cnt@rev.desc.d1.sspro.p1.C.grid2-d3.tau-source.ci95-excludes-0\endcsname{1}
\expandafter\def\csname pct@rev.desc.d1.sspro.p1.C.grid2-d3.tau-source.ci95-excludes-0\endcsname{100.0}
\expandafter\def\csname ppt@rev.desc.d1.sspro.p1.C.grid2-d3.tau-source.ci95-excludes-0\endcsname{$+$100.0}
\expandafter\def\csname val@rev.desc.d1.sspro.p1.C.grid2-d3.tau-source.ci95.0\endcsname{0.062}
\expandafter\def\csname cnt@rev.desc.d1.sspro.p1.C.grid2-d3.tau-source.ci95.0\endcsname{0}
\expandafter\def\csname pct@rev.desc.d1.sspro.p1.C.grid2-d3.tau-source.ci95.0\endcsname{6.2}
\expandafter\def\csname ppt@rev.desc.d1.sspro.p1.C.grid2-d3.tau-source.ci95.0\endcsname{$+$6.2}
\expandafter\def\csname val@rev.desc.d1.sspro.p1.C.grid2-d3.tau-source.ci95.1\endcsname{0.354}
\expandafter\def\csname cnt@rev.desc.d1.sspro.p1.C.grid2-d3.tau-source.ci95.1\endcsname{0}
\expandafter\def\csname pct@rev.desc.d1.sspro.p1.C.grid2-d3.tau-source.ci95.1\endcsname{35.4}
\expandafter\def\csname ppt@rev.desc.d1.sspro.p1.C.grid2-d3.tau-source.ci95.1\endcsname{$+$35.4}
\expandafter\def\csname val@rev.desc.d1.sspro.p1.C.grid2-d3.tau-source.mean\endcsname{0.208}
\expandafter\def\csname cnt@rev.desc.d1.sspro.p1.C.grid2-d3.tau-source.mean\endcsname{0}
\expandafter\def\csname pct@rev.desc.d1.sspro.p1.C.grid2-d3.tau-source.mean\endcsname{20.8}
\expandafter\def\csname ppt@rev.desc.d1.sspro.p1.C.grid2-d3.tau-source.mean\endcsname{$+$20.8}
\expandafter\def\csname val@rev.desc.d1.sspro.p1.C.grid2-d3.tau-source.n-apps\endcsname{6}
\expandafter\def\csname cnt@rev.desc.d1.sspro.p1.C.grid2-d3.tau-source.n-apps\endcsname{6}
\expandafter\def\csname pct@rev.desc.d1.sspro.p1.C.grid2-d3.tau-source.n-apps\endcsname{600.0}
\expandafter\def\csname ppt@rev.desc.d1.sspro.p1.C.grid2-d3.tau-source.n-apps\endcsname{$+$600.0}
\expandafter\def\csname val@rev.desc.d1.sspro.p1.C.grid2-d3.tau-source.tost1.ci90-inside\endcsname{0}
\expandafter\def\csname cnt@rev.desc.d1.sspro.p1.C.grid2-d3.tau-source.tost1.ci90-inside\endcsname{0}
\expandafter\def\csname pct@rev.desc.d1.sspro.p1.C.grid2-d3.tau-source.tost1.ci90-inside\endcsname{0.0}
\expandafter\def\csname ppt@rev.desc.d1.sspro.p1.C.grid2-d3.tau-source.tost1.ci90-inside\endcsname{0.0}
\expandafter\def\csname val@rev.desc.d1.sspro.p1.C.grid2-d3.tau-source.tost1.equivalent\endcsname{0}
\expandafter\def\csname cnt@rev.desc.d1.sspro.p1.C.grid2-d3.tau-source.tost1.equivalent\endcsname{0}
\expandafter\def\csname pct@rev.desc.d1.sspro.p1.C.grid2-d3.tau-source.tost1.equivalent\endcsname{0.0}
\expandafter\def\csname ppt@rev.desc.d1.sspro.p1.C.grid2-d3.tau-source.tost1.equivalent\endcsname{0.0}
\expandafter\def\csname val@rev.desc.d1.sspro.p1.C.grid2-d3.tau-source.tost1.p\endcsname{0.999}
\expandafter\def\csname cnt@rev.desc.d1.sspro.p1.C.grid2-d3.tau-source.tost1.p\endcsname{1}
\expandafter\def\csname pct@rev.desc.d1.sspro.p1.C.grid2-d3.tau-source.tost1.p\endcsname{99.9}
\expandafter\def\csname ppt@rev.desc.d1.sspro.p1.C.grid2-d3.tau-source.tost1.p\endcsname{$+$99.9}
\expandafter\def\csname val@rev.desc.d1.sspro.p1.C.grid2-d3.tau-source.tost5.ci90-inside\endcsname{0}
\expandafter\def\csname cnt@rev.desc.d1.sspro.p1.C.grid2-d3.tau-source.tost5.ci90-inside\endcsname{0}
\expandafter\def\csname pct@rev.desc.d1.sspro.p1.C.grid2-d3.tau-source.tost5.ci90-inside\endcsname{0.0}
\expandafter\def\csname ppt@rev.desc.d1.sspro.p1.C.grid2-d3.tau-source.tost5.ci90-inside\endcsname{0.0}
\expandafter\def\csname val@rev.desc.d1.sspro.p1.C.grid2-d3.tau-source.tost5.equivalent\endcsname{0}
\expandafter\def\csname cnt@rev.desc.d1.sspro.p1.C.grid2-d3.tau-source.tost5.equivalent\endcsname{0}
\expandafter\def\csname pct@rev.desc.d1.sspro.p1.C.grid2-d3.tau-source.tost5.equivalent\endcsname{0.0}
\expandafter\def\csname ppt@rev.desc.d1.sspro.p1.C.grid2-d3.tau-source.tost5.equivalent\endcsname{0.0}
\expandafter\def\csname val@rev.desc.d1.sspro.p1.C.grid2-d3.tau-source.tost5.p\endcsname{0.987}
\expandafter\def\csname cnt@rev.desc.d1.sspro.p1.C.grid2-d3.tau-source.tost5.p\endcsname{1}
\expandafter\def\csname pct@rev.desc.d1.sspro.p1.C.grid2-d3.tau-source.tost5.p\endcsname{98.7}
\expandafter\def\csname ppt@rev.desc.d1.sspro.p1.C.grid2-d3.tau-source.tost5.p\endcsname{$+$98.7}
\expandafter\def\csname val@rev.desc.d1.sspro.p1.C.grid2-d3.tau-target.ci90.0\endcsname{0.062}
\expandafter\def\csname cnt@rev.desc.d1.sspro.p1.C.grid2-d3.tau-target.ci90.0\endcsname{0}
\expandafter\def\csname pct@rev.desc.d1.sspro.p1.C.grid2-d3.tau-target.ci90.0\endcsname{6.2}
\expandafter\def\csname ppt@rev.desc.d1.sspro.p1.C.grid2-d3.tau-target.ci90.0\endcsname{$+$6.2}
\expandafter\def\csname val@rev.desc.d1.sspro.p1.C.grid2-d3.tau-target.ci90.1\endcsname{0.250}
\expandafter\def\csname cnt@rev.desc.d1.sspro.p1.C.grid2-d3.tau-target.ci90.1\endcsname{0}
\expandafter\def\csname pct@rev.desc.d1.sspro.p1.C.grid2-d3.tau-target.ci90.1\endcsname{25.0}
\expandafter\def\csname ppt@rev.desc.d1.sspro.p1.C.grid2-d3.tau-target.ci90.1\endcsname{$+$25.0}
\expandafter\def\csname val@rev.desc.d1.sspro.p1.C.grid2-d3.tau-target.ci95-excludes-0\endcsname{1}
\expandafter\def\csname cnt@rev.desc.d1.sspro.p1.C.grid2-d3.tau-target.ci95-excludes-0\endcsname{1}
\expandafter\def\csname pct@rev.desc.d1.sspro.p1.C.grid2-d3.tau-target.ci95-excludes-0\endcsname{100.0}
\expandafter\def\csname ppt@rev.desc.d1.sspro.p1.C.grid2-d3.tau-target.ci95-excludes-0\endcsname{$+$100.0}
\expandafter\def\csname val@rev.desc.d1.sspro.p1.C.grid2-d3.tau-target.ci95.0\endcsname{0.042}
\expandafter\def\csname cnt@rev.desc.d1.sspro.p1.C.grid2-d3.tau-target.ci95.0\endcsname{0}
\expandafter\def\csname pct@rev.desc.d1.sspro.p1.C.grid2-d3.tau-target.ci95.0\endcsname{4.2}
\expandafter\def\csname ppt@rev.desc.d1.sspro.p1.C.grid2-d3.tau-target.ci95.0\endcsname{$+$4.2}
\expandafter\def\csname val@rev.desc.d1.sspro.p1.C.grid2-d3.tau-target.ci95.1\endcsname{0.250}
\expandafter\def\csname cnt@rev.desc.d1.sspro.p1.C.grid2-d3.tau-target.ci95.1\endcsname{0}
\expandafter\def\csname pct@rev.desc.d1.sspro.p1.C.grid2-d3.tau-target.ci95.1\endcsname{25.0}
\expandafter\def\csname ppt@rev.desc.d1.sspro.p1.C.grid2-d3.tau-target.ci95.1\endcsname{$+$25.0}
\expandafter\def\csname val@rev.desc.d1.sspro.p1.C.grid2-d3.tau-target.mean\endcsname{0.146}
\expandafter\def\csname cnt@rev.desc.d1.sspro.p1.C.grid2-d3.tau-target.mean\endcsname{0}
\expandafter\def\csname pct@rev.desc.d1.sspro.p1.C.grid2-d3.tau-target.mean\endcsname{14.6}
\expandafter\def\csname ppt@rev.desc.d1.sspro.p1.C.grid2-d3.tau-target.mean\endcsname{$+$14.6}
\expandafter\def\csname val@rev.desc.d1.sspro.p1.C.grid2-d3.tau-target.n-apps\endcsname{6}
\expandafter\def\csname cnt@rev.desc.d1.sspro.p1.C.grid2-d3.tau-target.n-apps\endcsname{6}
\expandafter\def\csname pct@rev.desc.d1.sspro.p1.C.grid2-d3.tau-target.n-apps\endcsname{600.0}
\expandafter\def\csname ppt@rev.desc.d1.sspro.p1.C.grid2-d3.tau-target.n-apps\endcsname{$+$600.0}
\expandafter\def\csname val@rev.desc.d1.sspro.p1.C.grid2-d3.tau-target.tost1.ci90-inside\endcsname{0}
\expandafter\def\csname cnt@rev.desc.d1.sspro.p1.C.grid2-d3.tau-target.tost1.ci90-inside\endcsname{0}
\expandafter\def\csname pct@rev.desc.d1.sspro.p1.C.grid2-d3.tau-target.tost1.ci90-inside\endcsname{0.0}
\expandafter\def\csname ppt@rev.desc.d1.sspro.p1.C.grid2-d3.tau-target.tost1.ci90-inside\endcsname{0.0}
\expandafter\def\csname val@rev.desc.d1.sspro.p1.C.grid2-d3.tau-target.tost1.equivalent\endcsname{0}
\expandafter\def\csname cnt@rev.desc.d1.sspro.p1.C.grid2-d3.tau-target.tost1.equivalent\endcsname{0}
\expandafter\def\csname pct@rev.desc.d1.sspro.p1.C.grid2-d3.tau-target.tost1.equivalent\endcsname{0.0}
\expandafter\def\csname ppt@rev.desc.d1.sspro.p1.C.grid2-d3.tau-target.tost1.equivalent\endcsname{0.0}
\expandafter\def\csname val@rev.desc.d1.sspro.p1.C.grid2-d3.tau-target.tost1.p\endcsname{0.999}
\expandafter\def\csname cnt@rev.desc.d1.sspro.p1.C.grid2-d3.tau-target.tost1.p\endcsname{1}
\expandafter\def\csname pct@rev.desc.d1.sspro.p1.C.grid2-d3.tau-target.tost1.p\endcsname{99.9}
\expandafter\def\csname ppt@rev.desc.d1.sspro.p1.C.grid2-d3.tau-target.tost1.p\endcsname{$+$99.9}
\expandafter\def\csname val@rev.desc.d1.sspro.p1.C.grid2-d3.tau-target.tost5.ci90-inside\endcsname{0}
\expandafter\def\csname cnt@rev.desc.d1.sspro.p1.C.grid2-d3.tau-target.tost5.ci90-inside\endcsname{0}
\expandafter\def\csname pct@rev.desc.d1.sspro.p1.C.grid2-d3.tau-target.tost5.ci90-inside\endcsname{0.0}
\expandafter\def\csname ppt@rev.desc.d1.sspro.p1.C.grid2-d3.tau-target.tost5.ci90-inside\endcsname{0.0}
\expandafter\def\csname val@rev.desc.d1.sspro.p1.C.grid2-d3.tau-target.tost5.equivalent\endcsname{0}
\expandafter\def\csname cnt@rev.desc.d1.sspro.p1.C.grid2-d3.tau-target.tost5.equivalent\endcsname{0}
\expandafter\def\csname pct@rev.desc.d1.sspro.p1.C.grid2-d3.tau-target.tost5.equivalent\endcsname{0.0}
\expandafter\def\csname ppt@rev.desc.d1.sspro.p1.C.grid2-d3.tau-target.tost5.equivalent\endcsname{0.0}
\expandafter\def\csname val@rev.desc.d1.sspro.p1.C.grid2-d3.tau-target.tost5.p\endcsname{0.963}
\expandafter\def\csname cnt@rev.desc.d1.sspro.p1.C.grid2-d3.tau-target.tost5.p\endcsname{1}
\expandafter\def\csname pct@rev.desc.d1.sspro.p1.C.grid2-d3.tau-target.tost5.p\endcsname{96.3}
\expandafter\def\csname ppt@rev.desc.d1.sspro.p1.C.grid2-d3.tau-target.tost5.p\endcsname{$+$96.3}
\expandafter\def\csname val@rev.desc.d1.sspro.p1.C.grid2-d3.zero-estimate\endcsname{0}
\expandafter\def\csname cnt@rev.desc.d1.sspro.p1.C.grid2-d3.zero-estimate\endcsname{0}
\expandafter\def\csname pct@rev.desc.d1.sspro.p1.C.grid2-d3.zero-estimate\endcsname{0.0}
\expandafter\def\csname ppt@rev.desc.d1.sspro.p1.C.grid2-d3.zero-estimate\endcsname{0.0}
\expandafter\def\csname val@rev.desc.d1.sspro.p1.C.zoomclick-d3.delta-tau.ci90.0\endcsname{-0.125}
\expandafter\def\csname cnt@rev.desc.d1.sspro.p1.C.zoomclick-d3.delta-tau.ci90.0\endcsname{0}
\expandafter\def\csname pct@rev.desc.d1.sspro.p1.C.zoomclick-d3.delta-tau.ci90.0\endcsname{$-$12.5}
\expandafter\def\csname ppt@rev.desc.d1.sspro.p1.C.zoomclick-d3.delta-tau.ci90.0\endcsname{$-$12.5}
\expandafter\def\csname val@rev.desc.d1.sspro.p1.C.zoomclick-d3.delta-tau.ci90.1\endcsname{0.042}
\expandafter\def\csname cnt@rev.desc.d1.sspro.p1.C.zoomclick-d3.delta-tau.ci90.1\endcsname{0}
\expandafter\def\csname pct@rev.desc.d1.sspro.p1.C.zoomclick-d3.delta-tau.ci90.1\endcsname{4.2}
\expandafter\def\csname ppt@rev.desc.d1.sspro.p1.C.zoomclick-d3.delta-tau.ci90.1\endcsname{$+$4.2}
\expandafter\def\csname val@rev.desc.d1.sspro.p1.C.zoomclick-d3.delta-tau.ci95-excludes-0\endcsname{0}
\expandafter\def\csname cnt@rev.desc.d1.sspro.p1.C.zoomclick-d3.delta-tau.ci95-excludes-0\endcsname{0}
\expandafter\def\csname pct@rev.desc.d1.sspro.p1.C.zoomclick-d3.delta-tau.ci95-excludes-0\endcsname{0.0}
\expandafter\def\csname ppt@rev.desc.d1.sspro.p1.C.zoomclick-d3.delta-tau.ci95-excludes-0\endcsname{0.0}
\expandafter\def\csname val@rev.desc.d1.sspro.p1.C.zoomclick-d3.delta-tau.ci95.0\endcsname{-0.146}
\expandafter\def\csname cnt@rev.desc.d1.sspro.p1.C.zoomclick-d3.delta-tau.ci95.0\endcsname{0}
\expandafter\def\csname pct@rev.desc.d1.sspro.p1.C.zoomclick-d3.delta-tau.ci95.0\endcsname{$-$14.6}
\expandafter\def\csname ppt@rev.desc.d1.sspro.p1.C.zoomclick-d3.delta-tau.ci95.0\endcsname{$-$14.6}
\expandafter\def\csname val@rev.desc.d1.sspro.p1.C.zoomclick-d3.delta-tau.ci95.1\endcsname{0.042}
\expandafter\def\csname cnt@rev.desc.d1.sspro.p1.C.zoomclick-d3.delta-tau.ci95.1\endcsname{0}
\expandafter\def\csname pct@rev.desc.d1.sspro.p1.C.zoomclick-d3.delta-tau.ci95.1\endcsname{4.2}
\expandafter\def\csname ppt@rev.desc.d1.sspro.p1.C.zoomclick-d3.delta-tau.ci95.1\endcsname{$+$4.2}
\expandafter\def\csname val@rev.desc.d1.sspro.p1.C.zoomclick-d3.delta-tau.mean\endcsname{-0.042}
\expandafter\def\csname cnt@rev.desc.d1.sspro.p1.C.zoomclick-d3.delta-tau.mean\endcsname{0}
\expandafter\def\csname pct@rev.desc.d1.sspro.p1.C.zoomclick-d3.delta-tau.mean\endcsname{$-$4.2}
\expandafter\def\csname ppt@rev.desc.d1.sspro.p1.C.zoomclick-d3.delta-tau.mean\endcsname{$-$4.2}
\expandafter\def\csname val@rev.desc.d1.sspro.p1.C.zoomclick-d3.delta-tau.n-apps\endcsname{6}
\expandafter\def\csname cnt@rev.desc.d1.sspro.p1.C.zoomclick-d3.delta-tau.n-apps\endcsname{6}
\expandafter\def\csname pct@rev.desc.d1.sspro.p1.C.zoomclick-d3.delta-tau.n-apps\endcsname{600.0}
\expandafter\def\csname ppt@rev.desc.d1.sspro.p1.C.zoomclick-d3.delta-tau.n-apps\endcsname{$+$600.0}
\expandafter\def\csname val@rev.desc.d1.sspro.p1.C.zoomclick-d3.delta-tau.tost1.ci90-inside\endcsname{0}
\expandafter\def\csname cnt@rev.desc.d1.sspro.p1.C.zoomclick-d3.delta-tau.tost1.ci90-inside\endcsname{0}
\expandafter\def\csname pct@rev.desc.d1.sspro.p1.C.zoomclick-d3.delta-tau.tost1.ci90-inside\endcsname{0.0}
\expandafter\def\csname ppt@rev.desc.d1.sspro.p1.C.zoomclick-d3.delta-tau.tost1.ci90-inside\endcsname{0.0}
\expandafter\def\csname val@rev.desc.d1.sspro.p1.C.zoomclick-d3.delta-tau.tost1.equivalent\endcsname{0}
\expandafter\def\csname cnt@rev.desc.d1.sspro.p1.C.zoomclick-d3.delta-tau.tost1.equivalent\endcsname{0}
\expandafter\def\csname pct@rev.desc.d1.sspro.p1.C.zoomclick-d3.delta-tau.tost1.equivalent\endcsname{0.0}
\expandafter\def\csname ppt@rev.desc.d1.sspro.p1.C.zoomclick-d3.delta-tau.tost1.equivalent\endcsname{0.0}
\expandafter\def\csname val@rev.desc.d1.sspro.p1.C.zoomclick-d3.delta-tau.tost1.p\endcsname{0.740}
\expandafter\def\csname cnt@rev.desc.d1.sspro.p1.C.zoomclick-d3.delta-tau.tost1.p\endcsname{1}
\expandafter\def\csname pct@rev.desc.d1.sspro.p1.C.zoomclick-d3.delta-tau.tost1.p\endcsname{74.0}
\expandafter\def\csname ppt@rev.desc.d1.sspro.p1.C.zoomclick-d3.delta-tau.tost1.p\endcsname{$+$74.0}
\expandafter\def\csname val@rev.desc.d1.sspro.p1.C.zoomclick-d3.delta-tau.tost5.ci90-inside\endcsname{0}
\expandafter\def\csname cnt@rev.desc.d1.sspro.p1.C.zoomclick-d3.delta-tau.tost5.ci90-inside\endcsname{0}
\expandafter\def\csname pct@rev.desc.d1.sspro.p1.C.zoomclick-d3.delta-tau.tost5.ci90-inside\endcsname{0.0}
\expandafter\def\csname ppt@rev.desc.d1.sspro.p1.C.zoomclick-d3.delta-tau.tost5.ci90-inside\endcsname{0.0}
\expandafter\def\csname val@rev.desc.d1.sspro.p1.C.zoomclick-d3.delta-tau.tost5.equivalent\endcsname{0}
\expandafter\def\csname cnt@rev.desc.d1.sspro.p1.C.zoomclick-d3.delta-tau.tost5.equivalent\endcsname{0}
\expandafter\def\csname pct@rev.desc.d1.sspro.p1.C.zoomclick-d3.delta-tau.tost5.equivalent\endcsname{0.0}
\expandafter\def\csname ppt@rev.desc.d1.sspro.p1.C.zoomclick-d3.delta-tau.tost5.equivalent\endcsname{0.0}
\expandafter\def\csname val@rev.desc.d1.sspro.p1.C.zoomclick-d3.delta-tau.tost5.p\endcsname{0.402}
\expandafter\def\csname cnt@rev.desc.d1.sspro.p1.C.zoomclick-d3.delta-tau.tost5.p\endcsname{0}
\expandafter\def\csname pct@rev.desc.d1.sspro.p1.C.zoomclick-d3.delta-tau.tost5.p\endcsname{40.2}
\expandafter\def\csname ppt@rev.desc.d1.sspro.p1.C.zoomclick-d3.delta-tau.tost5.p\endcsname{$+$40.2}
\expandafter\def\csname val@rev.desc.d1.sspro.p1.C.zoomclick-d3.material-flip\endcsname{0}
\expandafter\def\csname cnt@rev.desc.d1.sspro.p1.C.zoomclick-d3.material-flip\endcsname{0}
\expandafter\def\csname pct@rev.desc.d1.sspro.p1.C.zoomclick-d3.material-flip\endcsname{0.0}
\expandafter\def\csname ppt@rev.desc.d1.sspro.p1.C.zoomclick-d3.material-flip\endcsname{0.0}
\expandafter\def\csname val@rev.desc.d1.sspro.p1.C.zoomclick-d3.n-apps\endcsname{6}
\expandafter\def\csname cnt@rev.desc.d1.sspro.p1.C.zoomclick-d3.n-apps\endcsname{6}
\expandafter\def\csname pct@rev.desc.d1.sspro.p1.C.zoomclick-d3.n-apps\endcsname{600.0}
\expandafter\def\csname ppt@rev.desc.d1.sspro.p1.C.zoomclick-d3.n-apps\endcsname{$+$600.0}
\expandafter\def\csname val@rev.desc.d1.sspro.p1.C.zoomclick-d3.n-tasks\endcsname{48}
\expandafter\def\csname cnt@rev.desc.d1.sspro.p1.C.zoomclick-d3.n-tasks\endcsname{48}
\expandafter\def\csname pct@rev.desc.d1.sspro.p1.C.zoomclick-d3.n-tasks\endcsname{4800.0}
\expandafter\def\csname ppt@rev.desc.d1.sspro.p1.C.zoomclick-d3.n-tasks\endcsname{$+$4800.0}
\expandafter\def\csname val@rev.desc.d1.sspro.p1.C.zoomclick-d3.sign-differs\endcsname{0}
\expandafter\def\csname cnt@rev.desc.d1.sspro.p1.C.zoomclick-d3.sign-differs\endcsname{0}
\expandafter\def\csname pct@rev.desc.d1.sspro.p1.C.zoomclick-d3.sign-differs\endcsname{0.0}
\expandafter\def\csname ppt@rev.desc.d1.sspro.p1.C.zoomclick-d3.sign-differs\endcsname{0.0}
\expandafter\def\csname val@rev.desc.d1.sspro.p1.C.zoomclick-d3.tau-source.ci90.0\endcsname{0.125}
\expandafter\def\csname cnt@rev.desc.d1.sspro.p1.C.zoomclick-d3.tau-source.ci90.0\endcsname{0}
\expandafter\def\csname pct@rev.desc.d1.sspro.p1.C.zoomclick-d3.tau-source.ci90.0\endcsname{12.5}
\expandafter\def\csname ppt@rev.desc.d1.sspro.p1.C.zoomclick-d3.tau-source.ci90.0\endcsname{$+$12.5}
\expandafter\def\csname val@rev.desc.d1.sspro.p1.C.zoomclick-d3.tau-source.ci90.1\endcsname{0.375}
\expandafter\def\csname cnt@rev.desc.d1.sspro.p1.C.zoomclick-d3.tau-source.ci90.1\endcsname{0}
\expandafter\def\csname pct@rev.desc.d1.sspro.p1.C.zoomclick-d3.tau-source.ci90.1\endcsname{37.5}
\expandafter\def\csname ppt@rev.desc.d1.sspro.p1.C.zoomclick-d3.tau-source.ci90.1\endcsname{$+$37.5}
\expandafter\def\csname val@rev.desc.d1.sspro.p1.C.zoomclick-d3.tau-source.ci95-excludes-0\endcsname{1}
\expandafter\def\csname cnt@rev.desc.d1.sspro.p1.C.zoomclick-d3.tau-source.ci95-excludes-0\endcsname{1}
\expandafter\def\csname pct@rev.desc.d1.sspro.p1.C.zoomclick-d3.tau-source.ci95-excludes-0\endcsname{100.0}
\expandafter\def\csname ppt@rev.desc.d1.sspro.p1.C.zoomclick-d3.tau-source.ci95-excludes-0\endcsname{$+$100.0}
\expandafter\def\csname val@rev.desc.d1.sspro.p1.C.zoomclick-d3.tau-source.ci95.0\endcsname{0.104}
\expandafter\def\csname cnt@rev.desc.d1.sspro.p1.C.zoomclick-d3.tau-source.ci95.0\endcsname{0}
\expandafter\def\csname pct@rev.desc.d1.sspro.p1.C.zoomclick-d3.tau-source.ci95.0\endcsname{10.4}
\expandafter\def\csname ppt@rev.desc.d1.sspro.p1.C.zoomclick-d3.tau-source.ci95.0\endcsname{$+$10.4}
\expandafter\def\csname val@rev.desc.d1.sspro.p1.C.zoomclick-d3.tau-source.ci95.1\endcsname{0.396}
\expandafter\def\csname cnt@rev.desc.d1.sspro.p1.C.zoomclick-d3.tau-source.ci95.1\endcsname{0}
\expandafter\def\csname pct@rev.desc.d1.sspro.p1.C.zoomclick-d3.tau-source.ci95.1\endcsname{39.6}
\expandafter\def\csname ppt@rev.desc.d1.sspro.p1.C.zoomclick-d3.tau-source.ci95.1\endcsname{$+$39.6}
\expandafter\def\csname val@rev.desc.d1.sspro.p1.C.zoomclick-d3.tau-source.mean\endcsname{0.250}
\expandafter\def\csname cnt@rev.desc.d1.sspro.p1.C.zoomclick-d3.tau-source.mean\endcsname{0}
\expandafter\def\csname pct@rev.desc.d1.sspro.p1.C.zoomclick-d3.tau-source.mean\endcsname{25.0}
\expandafter\def\csname ppt@rev.desc.d1.sspro.p1.C.zoomclick-d3.tau-source.mean\endcsname{$+$25.0}
\expandafter\def\csname val@rev.desc.d1.sspro.p1.C.zoomclick-d3.tau-source.n-apps\endcsname{6}
\expandafter\def\csname cnt@rev.desc.d1.sspro.p1.C.zoomclick-d3.tau-source.n-apps\endcsname{6}
\expandafter\def\csname pct@rev.desc.d1.sspro.p1.C.zoomclick-d3.tau-source.n-apps\endcsname{600.0}
\expandafter\def\csname ppt@rev.desc.d1.sspro.p1.C.zoomclick-d3.tau-source.n-apps\endcsname{$+$600.0}
\expandafter\def\csname val@rev.desc.d1.sspro.p1.C.zoomclick-d3.tau-source.tost1.ci90-inside\endcsname{0}
\expandafter\def\csname cnt@rev.desc.d1.sspro.p1.C.zoomclick-d3.tau-source.tost1.ci90-inside\endcsname{0}
\expandafter\def\csname pct@rev.desc.d1.sspro.p1.C.zoomclick-d3.tau-source.tost1.ci90-inside\endcsname{0.0}
\expandafter\def\csname ppt@rev.desc.d1.sspro.p1.C.zoomclick-d3.tau-source.tost1.ci90-inside\endcsname{0.0}
\expandafter\def\csname val@rev.desc.d1.sspro.p1.C.zoomclick-d3.tau-source.tost1.equivalent\endcsname{0}
\expandafter\def\csname cnt@rev.desc.d1.sspro.p1.C.zoomclick-d3.tau-source.tost1.equivalent\endcsname{0}
\expandafter\def\csname pct@rev.desc.d1.sspro.p1.C.zoomclick-d3.tau-source.tost1.equivalent\endcsname{0.0}
\expandafter\def\csname ppt@rev.desc.d1.sspro.p1.C.zoomclick-d3.tau-source.tost1.equivalent\endcsname{0.0}
\expandafter\def\csname val@rev.desc.d1.sspro.p1.C.zoomclick-d3.tau-source.tost1.p\endcsname{1.000}
\expandafter\def\csname cnt@rev.desc.d1.sspro.p1.C.zoomclick-d3.tau-source.tost1.p\endcsname{1}
\expandafter\def\csname pct@rev.desc.d1.sspro.p1.C.zoomclick-d3.tau-source.tost1.p\endcsname{100.0}
\expandafter\def\csname ppt@rev.desc.d1.sspro.p1.C.zoomclick-d3.tau-source.tost1.p\endcsname{$+$100.0}
\expandafter\def\csname val@rev.desc.d1.sspro.p1.C.zoomclick-d3.tau-source.tost5.ci90-inside\endcsname{0}
\expandafter\def\csname cnt@rev.desc.d1.sspro.p1.C.zoomclick-d3.tau-source.tost5.ci90-inside\endcsname{0}
\expandafter\def\csname pct@rev.desc.d1.sspro.p1.C.zoomclick-d3.tau-source.tost5.ci90-inside\endcsname{0.0}
\expandafter\def\csname ppt@rev.desc.d1.sspro.p1.C.zoomclick-d3.tau-source.tost5.ci90-inside\endcsname{0.0}
\expandafter\def\csname val@rev.desc.d1.sspro.p1.C.zoomclick-d3.tau-source.tost5.equivalent\endcsname{0}
\expandafter\def\csname cnt@rev.desc.d1.sspro.p1.C.zoomclick-d3.tau-source.tost5.equivalent\endcsname{0}
\expandafter\def\csname pct@rev.desc.d1.sspro.p1.C.zoomclick-d3.tau-source.tost5.equivalent\endcsname{0.0}
\expandafter\def\csname ppt@rev.desc.d1.sspro.p1.C.zoomclick-d3.tau-source.tost5.equivalent\endcsname{0.0}
\expandafter\def\csname val@rev.desc.d1.sspro.p1.C.zoomclick-d3.tau-source.tost5.p\endcsname{0.998}
\expandafter\def\csname cnt@rev.desc.d1.sspro.p1.C.zoomclick-d3.tau-source.tost5.p\endcsname{1}
\expandafter\def\csname pct@rev.desc.d1.sspro.p1.C.zoomclick-d3.tau-source.tost5.p\endcsname{99.8}
\expandafter\def\csname ppt@rev.desc.d1.sspro.p1.C.zoomclick-d3.tau-source.tost5.p\endcsname{$+$99.8}
\expandafter\def\csname val@rev.desc.d1.sspro.p1.C.zoomclick-d3.tau-target.ci90.0\endcsname{0.125}
\expandafter\def\csname cnt@rev.desc.d1.sspro.p1.C.zoomclick-d3.tau-target.ci90.0\endcsname{0}
\expandafter\def\csname pct@rev.desc.d1.sspro.p1.C.zoomclick-d3.tau-target.ci90.0\endcsname{12.5}
\expandafter\def\csname ppt@rev.desc.d1.sspro.p1.C.zoomclick-d3.tau-target.ci90.0\endcsname{$+$12.5}
\expandafter\def\csname val@rev.desc.d1.sspro.p1.C.zoomclick-d3.tau-target.ci90.1\endcsname{0.292}
\expandafter\def\csname cnt@rev.desc.d1.sspro.p1.C.zoomclick-d3.tau-target.ci90.1\endcsname{0}
\expandafter\def\csname pct@rev.desc.d1.sspro.p1.C.zoomclick-d3.tau-target.ci90.1\endcsname{29.2}
\expandafter\def\csname ppt@rev.desc.d1.sspro.p1.C.zoomclick-d3.tau-target.ci90.1\endcsname{$+$29.2}
\expandafter\def\csname val@rev.desc.d1.sspro.p1.C.zoomclick-d3.tau-target.ci95-excludes-0\endcsname{1}
\expandafter\def\csname cnt@rev.desc.d1.sspro.p1.C.zoomclick-d3.tau-target.ci95-excludes-0\endcsname{1}
\expandafter\def\csname pct@rev.desc.d1.sspro.p1.C.zoomclick-d3.tau-target.ci95-excludes-0\endcsname{100.0}
\expandafter\def\csname ppt@rev.desc.d1.sspro.p1.C.zoomclick-d3.tau-target.ci95-excludes-0\endcsname{$+$100.0}
\expandafter\def\csname val@rev.desc.d1.sspro.p1.C.zoomclick-d3.tau-target.ci95.0\endcsname{0.125}
\expandafter\def\csname cnt@rev.desc.d1.sspro.p1.C.zoomclick-d3.tau-target.ci95.0\endcsname{0}
\expandafter\def\csname pct@rev.desc.d1.sspro.p1.C.zoomclick-d3.tau-target.ci95.0\endcsname{12.5}
\expandafter\def\csname ppt@rev.desc.d1.sspro.p1.C.zoomclick-d3.tau-target.ci95.0\endcsname{$+$12.5}
\expandafter\def\csname val@rev.desc.d1.sspro.p1.C.zoomclick-d3.tau-target.ci95.1\endcsname{0.292}
\expandafter\def\csname cnt@rev.desc.d1.sspro.p1.C.zoomclick-d3.tau-target.ci95.1\endcsname{0}
\expandafter\def\csname pct@rev.desc.d1.sspro.p1.C.zoomclick-d3.tau-target.ci95.1\endcsname{29.2}
\expandafter\def\csname ppt@rev.desc.d1.sspro.p1.C.zoomclick-d3.tau-target.ci95.1\endcsname{$+$29.2}
\expandafter\def\csname val@rev.desc.d1.sspro.p1.C.zoomclick-d3.tau-target.mean\endcsname{0.208}
\expandafter\def\csname cnt@rev.desc.d1.sspro.p1.C.zoomclick-d3.tau-target.mean\endcsname{0}
\expandafter\def\csname pct@rev.desc.d1.sspro.p1.C.zoomclick-d3.tau-target.mean\endcsname{20.8}
\expandafter\def\csname ppt@rev.desc.d1.sspro.p1.C.zoomclick-d3.tau-target.mean\endcsname{$+$20.8}
\expandafter\def\csname val@rev.desc.d1.sspro.p1.C.zoomclick-d3.tau-target.n-apps\endcsname{6}
\expandafter\def\csname cnt@rev.desc.d1.sspro.p1.C.zoomclick-d3.tau-target.n-apps\endcsname{6}
\expandafter\def\csname pct@rev.desc.d1.sspro.p1.C.zoomclick-d3.tau-target.n-apps\endcsname{600.0}
\expandafter\def\csname ppt@rev.desc.d1.sspro.p1.C.zoomclick-d3.tau-target.n-apps\endcsname{$+$600.0}
\expandafter\def\csname val@rev.desc.d1.sspro.p1.C.zoomclick-d3.tau-target.tost1.ci90-inside\endcsname{0}
\expandafter\def\csname cnt@rev.desc.d1.sspro.p1.C.zoomclick-d3.tau-target.tost1.ci90-inside\endcsname{0}
\expandafter\def\csname pct@rev.desc.d1.sspro.p1.C.zoomclick-d3.tau-target.tost1.ci90-inside\endcsname{0.0}
\expandafter\def\csname ppt@rev.desc.d1.sspro.p1.C.zoomclick-d3.tau-target.tost1.ci90-inside\endcsname{0.0}
\expandafter\def\csname val@rev.desc.d1.sspro.p1.C.zoomclick-d3.tau-target.tost1.equivalent\endcsname{0}
\expandafter\def\csname cnt@rev.desc.d1.sspro.p1.C.zoomclick-d3.tau-target.tost1.equivalent\endcsname{0}
\expandafter\def\csname pct@rev.desc.d1.sspro.p1.C.zoomclick-d3.tau-target.tost1.equivalent\endcsname{0.0}
\expandafter\def\csname ppt@rev.desc.d1.sspro.p1.C.zoomclick-d3.tau-target.tost1.equivalent\endcsname{0.0}
\expandafter\def\csname val@rev.desc.d1.sspro.p1.C.zoomclick-d3.tau-target.tost1.p\endcsname{1.000}
\expandafter\def\csname cnt@rev.desc.d1.sspro.p1.C.zoomclick-d3.tau-target.tost1.p\endcsname{1}
\expandafter\def\csname pct@rev.desc.d1.sspro.p1.C.zoomclick-d3.tau-target.tost1.p\endcsname{100.0}
\expandafter\def\csname ppt@rev.desc.d1.sspro.p1.C.zoomclick-d3.tau-target.tost1.p\endcsname{$+$100.0}
\expandafter\def\csname val@rev.desc.d1.sspro.p1.C.zoomclick-d3.tau-target.tost5.ci90-inside\endcsname{0}
\expandafter\def\csname cnt@rev.desc.d1.sspro.p1.C.zoomclick-d3.tau-target.tost5.ci90-inside\endcsname{0}
\expandafter\def\csname pct@rev.desc.d1.sspro.p1.C.zoomclick-d3.tau-target.tost5.ci90-inside\endcsname{0.0}
\expandafter\def\csname ppt@rev.desc.d1.sspro.p1.C.zoomclick-d3.tau-target.tost5.ci90-inside\endcsname{0.0}
\expandafter\def\csname val@rev.desc.d1.sspro.p1.C.zoomclick-d3.tau-target.tost5.equivalent\endcsname{0}
\expandafter\def\csname cnt@rev.desc.d1.sspro.p1.C.zoomclick-d3.tau-target.tost5.equivalent\endcsname{0}
\expandafter\def\csname pct@rev.desc.d1.sspro.p1.C.zoomclick-d3.tau-target.tost5.equivalent\endcsname{0.0}
\expandafter\def\csname ppt@rev.desc.d1.sspro.p1.C.zoomclick-d3.tau-target.tost5.equivalent\endcsname{0.0}
\expandafter\def\csname val@rev.desc.d1.sspro.p1.C.zoomclick-d3.tau-target.tost5.p\endcsname{1.000}
\expandafter\def\csname cnt@rev.desc.d1.sspro.p1.C.zoomclick-d3.tau-target.tost5.p\endcsname{1}
\expandafter\def\csname pct@rev.desc.d1.sspro.p1.C.zoomclick-d3.tau-target.tost5.p\endcsname{100.0}
\expandafter\def\csname ppt@rev.desc.d1.sspro.p1.C.zoomclick-d3.tau-target.tost5.p\endcsname{$+$100.0}
\expandafter\def\csname val@rev.desc.d1.sspro.p1.C.zoomclick-d3.zero-estimate\endcsname{0}
\expandafter\def\csname cnt@rev.desc.d1.sspro.p1.C.zoomclick-d3.zero-estimate\endcsname{0}
\expandafter\def\csname pct@rev.desc.d1.sspro.p1.C.zoomclick-d3.zero-estimate\endcsname{0.0}
\expandafter\def\csname ppt@rev.desc.d1.sspro.p1.C.zoomclick-d3.zero-estimate\endcsname{0.0}
\expandafter\def\csname val@rev.desc.d1.sspro.p1.Tsealed.center-d1.delta-tau-cross.ci90.0\endcsname{-0.188}
\expandafter\def\csname cnt@rev.desc.d1.sspro.p1.Tsealed.center-d1.delta-tau-cross.ci90.0\endcsname{0}
\expandafter\def\csname pct@rev.desc.d1.sspro.p1.Tsealed.center-d1.delta-tau-cross.ci90.0\endcsname{$-$18.8}
\expandafter\def\csname ppt@rev.desc.d1.sspro.p1.Tsealed.center-d1.delta-tau-cross.ci90.0\endcsname{$-$18.8}
\expandafter\def\csname val@rev.desc.d1.sspro.p1.Tsealed.center-d1.delta-tau-cross.ci90.1\endcsname{0.094}
\expandafter\def\csname cnt@rev.desc.d1.sspro.p1.Tsealed.center-d1.delta-tau-cross.ci90.1\endcsname{0}
\expandafter\def\csname pct@rev.desc.d1.sspro.p1.Tsealed.center-d1.delta-tau-cross.ci90.1\endcsname{9.4}
\expandafter\def\csname ppt@rev.desc.d1.sspro.p1.Tsealed.center-d1.delta-tau-cross.ci90.1\endcsname{$+$9.4}
\expandafter\def\csname val@rev.desc.d1.sspro.p1.Tsealed.center-d1.delta-tau-cross.ci95-excludes-0\endcsname{0}
\expandafter\def\csname cnt@rev.desc.d1.sspro.p1.Tsealed.center-d1.delta-tau-cross.ci95-excludes-0\endcsname{0}
\expandafter\def\csname pct@rev.desc.d1.sspro.p1.Tsealed.center-d1.delta-tau-cross.ci95-excludes-0\endcsname{0.0}
\expandafter\def\csname ppt@rev.desc.d1.sspro.p1.Tsealed.center-d1.delta-tau-cross.ci95-excludes-0\endcsname{0.0}
\expandafter\def\csname val@rev.desc.d1.sspro.p1.Tsealed.center-d1.delta-tau-cross.ci95.0\endcsname{-0.219}
\expandafter\def\csname cnt@rev.desc.d1.sspro.p1.Tsealed.center-d1.delta-tau-cross.ci95.0\endcsname{0}
\expandafter\def\csname pct@rev.desc.d1.sspro.p1.Tsealed.center-d1.delta-tau-cross.ci95.0\endcsname{$-$21.9}
\expandafter\def\csname ppt@rev.desc.d1.sspro.p1.Tsealed.center-d1.delta-tau-cross.ci95.0\endcsname{$-$21.9}
\expandafter\def\csname val@rev.desc.d1.sspro.p1.Tsealed.center-d1.delta-tau-cross.ci95.1\endcsname{0.115}
\expandafter\def\csname cnt@rev.desc.d1.sspro.p1.Tsealed.center-d1.delta-tau-cross.ci95.1\endcsname{0}
\expandafter\def\csname pct@rev.desc.d1.sspro.p1.Tsealed.center-d1.delta-tau-cross.ci95.1\endcsname{11.5}
\expandafter\def\csname ppt@rev.desc.d1.sspro.p1.Tsealed.center-d1.delta-tau-cross.ci95.1\endcsname{$+$11.5}
\expandafter\def\csname val@rev.desc.d1.sspro.p1.Tsealed.center-d1.delta-tau-cross.mean\endcsname{-0.042}
\expandafter\def\csname cnt@rev.desc.d1.sspro.p1.Tsealed.center-d1.delta-tau-cross.mean\endcsname{0}
\expandafter\def\csname pct@rev.desc.d1.sspro.p1.Tsealed.center-d1.delta-tau-cross.mean\endcsname{$-$4.2}
\expandafter\def\csname ppt@rev.desc.d1.sspro.p1.Tsealed.center-d1.delta-tau-cross.mean\endcsname{$-$4.2}
\expandafter\def\csname val@rev.desc.d1.sspro.p1.Tsealed.center-d1.delta-tau-cross.n-apps.0\endcsname{12}
\expandafter\def\csname cnt@rev.desc.d1.sspro.p1.Tsealed.center-d1.delta-tau-cross.n-apps.0\endcsname{12}
\expandafter\def\csname pct@rev.desc.d1.sspro.p1.Tsealed.center-d1.delta-tau-cross.n-apps.0\endcsname{1200.0}
\expandafter\def\csname ppt@rev.desc.d1.sspro.p1.Tsealed.center-d1.delta-tau-cross.n-apps.0\endcsname{$+$1200.0}
\expandafter\def\csname val@rev.desc.d1.sspro.p1.Tsealed.center-d1.delta-tau-cross.n-apps.1\endcsname{6}
\expandafter\def\csname cnt@rev.desc.d1.sspro.p1.Tsealed.center-d1.delta-tau-cross.n-apps.1\endcsname{6}
\expandafter\def\csname pct@rev.desc.d1.sspro.p1.Tsealed.center-d1.delta-tau-cross.n-apps.1\endcsname{600.0}
\expandafter\def\csname ppt@rev.desc.d1.sspro.p1.Tsealed.center-d1.delta-tau-cross.n-apps.1\endcsname{$+$600.0}
\expandafter\def\csname val@rev.desc.d1.sspro.p1.Tsealed.center-d1.delta-tau-cross.tost1.ci90-inside\endcsname{0}
\expandafter\def\csname cnt@rev.desc.d1.sspro.p1.Tsealed.center-d1.delta-tau-cross.tost1.ci90-inside\endcsname{0}
\expandafter\def\csname pct@rev.desc.d1.sspro.p1.Tsealed.center-d1.delta-tau-cross.tost1.ci90-inside\endcsname{0.0}
\expandafter\def\csname ppt@rev.desc.d1.sspro.p1.Tsealed.center-d1.delta-tau-cross.tost1.ci90-inside\endcsname{0.0}
\expandafter\def\csname val@rev.desc.d1.sspro.p1.Tsealed.center-d1.delta-tau-cross.tost1.equivalent\endcsname{0}
\expandafter\def\csname cnt@rev.desc.d1.sspro.p1.Tsealed.center-d1.delta-tau-cross.tost1.equivalent\endcsname{0}
\expandafter\def\csname pct@rev.desc.d1.sspro.p1.Tsealed.center-d1.delta-tau-cross.tost1.equivalent\endcsname{0.0}
\expandafter\def\csname ppt@rev.desc.d1.sspro.p1.Tsealed.center-d1.delta-tau-cross.tost1.equivalent\endcsname{0.0}
\expandafter\def\csname val@rev.desc.d1.sspro.p1.Tsealed.center-d1.delta-tau-cross.tost1.p\endcsname{0.650}
\expandafter\def\csname cnt@rev.desc.d1.sspro.p1.Tsealed.center-d1.delta-tau-cross.tost1.p\endcsname{1}
\expandafter\def\csname pct@rev.desc.d1.sspro.p1.Tsealed.center-d1.delta-tau-cross.tost1.p\endcsname{65.0}
\expandafter\def\csname ppt@rev.desc.d1.sspro.p1.Tsealed.center-d1.delta-tau-cross.tost1.p\endcsname{$+$65.0}
\expandafter\def\csname val@rev.desc.d1.sspro.p1.Tsealed.center-d1.delta-tau-cross.tost5.ci90-inside\endcsname{0}
\expandafter\def\csname cnt@rev.desc.d1.sspro.p1.Tsealed.center-d1.delta-tau-cross.tost5.ci90-inside\endcsname{0}
\expandafter\def\csname pct@rev.desc.d1.sspro.p1.Tsealed.center-d1.delta-tau-cross.tost5.ci90-inside\endcsname{0.0}
\expandafter\def\csname ppt@rev.desc.d1.sspro.p1.Tsealed.center-d1.delta-tau-cross.tost5.ci90-inside\endcsname{0.0}
\expandafter\def\csname val@rev.desc.d1.sspro.p1.Tsealed.center-d1.delta-tau-cross.tost5.equivalent\endcsname{0}
\expandafter\def\csname cnt@rev.desc.d1.sspro.p1.Tsealed.center-d1.delta-tau-cross.tost5.equivalent\endcsname{0}
\expandafter\def\csname pct@rev.desc.d1.sspro.p1.Tsealed.center-d1.delta-tau-cross.tost5.equivalent\endcsname{0.0}
\expandafter\def\csname ppt@rev.desc.d1.sspro.p1.Tsealed.center-d1.delta-tau-cross.tost5.equivalent\endcsname{0.0}
\expandafter\def\csname val@rev.desc.d1.sspro.p1.Tsealed.center-d1.delta-tau-cross.tost5.p\endcsname{0.462}
\expandafter\def\csname cnt@rev.desc.d1.sspro.p1.Tsealed.center-d1.delta-tau-cross.tost5.p\endcsname{0}
\expandafter\def\csname pct@rev.desc.d1.sspro.p1.Tsealed.center-d1.delta-tau-cross.tost5.p\endcsname{46.2}
\expandafter\def\csname ppt@rev.desc.d1.sspro.p1.Tsealed.center-d1.delta-tau-cross.tost5.p\endcsname{$+$46.2}
\expandafter\def\csname val@rev.desc.d1.sspro.p1.Tsealed.center-d1.material-flip\endcsname{0}
\expandafter\def\csname cnt@rev.desc.d1.sspro.p1.Tsealed.center-d1.material-flip\endcsname{0}
\expandafter\def\csname pct@rev.desc.d1.sspro.p1.Tsealed.center-d1.material-flip\endcsname{0.0}
\expandafter\def\csname ppt@rev.desc.d1.sspro.p1.Tsealed.center-d1.material-flip\endcsname{0.0}
\expandafter\def\csname val@rev.desc.d1.sspro.p1.Tsealed.center-d1.n-tasks\endcsname{96}
\expandafter\def\csname cnt@rev.desc.d1.sspro.p1.Tsealed.center-d1.n-tasks\endcsname{96}
\expandafter\def\csname pct@rev.desc.d1.sspro.p1.Tsealed.center-d1.n-tasks\endcsname{9600.0}
\expandafter\def\csname ppt@rev.desc.d1.sspro.p1.Tsealed.center-d1.n-tasks\endcsname{$+$9600.0}
\expandafter\def\csname val@rev.desc.d1.sspro.p1.Tsealed.center-d1.n-tasks-source-C\endcsname{48}
\expandafter\def\csname cnt@rev.desc.d1.sspro.p1.Tsealed.center-d1.n-tasks-source-C\endcsname{48}
\expandafter\def\csname pct@rev.desc.d1.sspro.p1.Tsealed.center-d1.n-tasks-source-C\endcsname{4800.0}
\expandafter\def\csname ppt@rev.desc.d1.sspro.p1.Tsealed.center-d1.n-tasks-source-C\endcsname{$+$4800.0}
\expandafter\def\csname val@rev.desc.d1.sspro.p1.Tsealed.center-d1.paper-tau-T\endcsname{0.104}
\expandafter\def\csname cnt@rev.desc.d1.sspro.p1.Tsealed.center-d1.paper-tau-T\endcsname{0}
\expandafter\def\csname pct@rev.desc.d1.sspro.p1.Tsealed.center-d1.paper-tau-T\endcsname{10.4}
\expandafter\def\csname ppt@rev.desc.d1.sspro.p1.Tsealed.center-d1.paper-tau-T\endcsname{$+$10.4}
\expandafter\def\csname val@rev.desc.d1.sspro.p1.Tsealed.center-d1.reproduces\endcsname{1}
\expandafter\def\csname cnt@rev.desc.d1.sspro.p1.Tsealed.center-d1.reproduces\endcsname{1}
\expandafter\def\csname pct@rev.desc.d1.sspro.p1.Tsealed.center-d1.reproduces\endcsname{100.0}
\expandafter\def\csname ppt@rev.desc.d1.sspro.p1.Tsealed.center-d1.reproduces\endcsname{$+$100.0}
\expandafter\def\csname val@rev.desc.d1.sspro.p1.Tsealed.center-d1.reproduces-ci\endcsname{1}
\expandafter\def\csname cnt@rev.desc.d1.sspro.p1.Tsealed.center-d1.reproduces-ci\endcsname{1}
\expandafter\def\csname pct@rev.desc.d1.sspro.p1.Tsealed.center-d1.reproduces-ci\endcsname{100.0}
\expandafter\def\csname ppt@rev.desc.d1.sspro.p1.Tsealed.center-d1.reproduces-ci\endcsname{$+$100.0}
\expandafter\def\csname val@rev.desc.d1.sspro.p1.Tsealed.center-d1.sign-differs\endcsname{0}
\expandafter\def\csname cnt@rev.desc.d1.sspro.p1.Tsealed.center-d1.sign-differs\endcsname{0}
\expandafter\def\csname pct@rev.desc.d1.sspro.p1.Tsealed.center-d1.sign-differs\endcsname{0.0}
\expandafter\def\csname ppt@rev.desc.d1.sspro.p1.Tsealed.center-d1.sign-differs\endcsname{0.0}
\expandafter\def\csname val@rev.desc.d1.sspro.p1.Tsealed.center-d1.tau-source-C.ci90.0\endcsname{0.042}
\expandafter\def\csname cnt@rev.desc.d1.sspro.p1.Tsealed.center-d1.tau-source-C.ci90.0\endcsname{0}
\expandafter\def\csname pct@rev.desc.d1.sspro.p1.Tsealed.center-d1.tau-source-C.ci90.0\endcsname{4.2}
\expandafter\def\csname ppt@rev.desc.d1.sspro.p1.Tsealed.center-d1.tau-source-C.ci90.0\endcsname{$+$4.2}
\expandafter\def\csname val@rev.desc.d1.sspro.p1.Tsealed.center-d1.tau-source-C.ci90.1\endcsname{0.271}
\expandafter\def\csname cnt@rev.desc.d1.sspro.p1.Tsealed.center-d1.tau-source-C.ci90.1\endcsname{0}
\expandafter\def\csname pct@rev.desc.d1.sspro.p1.Tsealed.center-d1.tau-source-C.ci90.1\endcsname{27.1}
\expandafter\def\csname ppt@rev.desc.d1.sspro.p1.Tsealed.center-d1.tau-source-C.ci90.1\endcsname{$+$27.1}
\expandafter\def\csname val@rev.desc.d1.sspro.p1.Tsealed.center-d1.tau-source-C.ci95-excludes-0\endcsname{1}
\expandafter\def\csname cnt@rev.desc.d1.sspro.p1.Tsealed.center-d1.tau-source-C.ci95-excludes-0\endcsname{1}
\expandafter\def\csname pct@rev.desc.d1.sspro.p1.Tsealed.center-d1.tau-source-C.ci95-excludes-0\endcsname{100.0}
\expandafter\def\csname ppt@rev.desc.d1.sspro.p1.Tsealed.center-d1.tau-source-C.ci95-excludes-0\endcsname{$+$100.0}
\expandafter\def\csname val@rev.desc.d1.sspro.p1.Tsealed.center-d1.tau-source-C.ci95.0\endcsname{0.042}
\expandafter\def\csname cnt@rev.desc.d1.sspro.p1.Tsealed.center-d1.tau-source-C.ci95.0\endcsname{0}
\expandafter\def\csname pct@rev.desc.d1.sspro.p1.Tsealed.center-d1.tau-source-C.ci95.0\endcsname{4.2}
\expandafter\def\csname ppt@rev.desc.d1.sspro.p1.Tsealed.center-d1.tau-source-C.ci95.0\endcsname{$+$4.2}
\expandafter\def\csname val@rev.desc.d1.sspro.p1.Tsealed.center-d1.tau-source-C.ci95.1\endcsname{0.292}
\expandafter\def\csname cnt@rev.desc.d1.sspro.p1.Tsealed.center-d1.tau-source-C.ci95.1\endcsname{0}
\expandafter\def\csname pct@rev.desc.d1.sspro.p1.Tsealed.center-d1.tau-source-C.ci95.1\endcsname{29.2}
\expandafter\def\csname ppt@rev.desc.d1.sspro.p1.Tsealed.center-d1.tau-source-C.ci95.1\endcsname{$+$29.2}
\expandafter\def\csname val@rev.desc.d1.sspro.p1.Tsealed.center-d1.tau-source-C.mean\endcsname{0.146}
\expandafter\def\csname cnt@rev.desc.d1.sspro.p1.Tsealed.center-d1.tau-source-C.mean\endcsname{0}
\expandafter\def\csname pct@rev.desc.d1.sspro.p1.Tsealed.center-d1.tau-source-C.mean\endcsname{14.6}
\expandafter\def\csname ppt@rev.desc.d1.sspro.p1.Tsealed.center-d1.tau-source-C.mean\endcsname{$+$14.6}
\expandafter\def\csname val@rev.desc.d1.sspro.p1.Tsealed.center-d1.tau-source-C.n-apps\endcsname{6}
\expandafter\def\csname cnt@rev.desc.d1.sspro.p1.Tsealed.center-d1.tau-source-C.n-apps\endcsname{6}
\expandafter\def\csname pct@rev.desc.d1.sspro.p1.Tsealed.center-d1.tau-source-C.n-apps\endcsname{600.0}
\expandafter\def\csname ppt@rev.desc.d1.sspro.p1.Tsealed.center-d1.tau-source-C.n-apps\endcsname{$+$600.0}
\expandafter\def\csname val@rev.desc.d1.sspro.p1.Tsealed.center-d1.tau-source-C.tost1.ci90-inside\endcsname{0}
\expandafter\def\csname cnt@rev.desc.d1.sspro.p1.Tsealed.center-d1.tau-source-C.tost1.ci90-inside\endcsname{0}
\expandafter\def\csname pct@rev.desc.d1.sspro.p1.Tsealed.center-d1.tau-source-C.tost1.ci90-inside\endcsname{0.0}
\expandafter\def\csname ppt@rev.desc.d1.sspro.p1.Tsealed.center-d1.tau-source-C.tost1.ci90-inside\endcsname{0.0}
\expandafter\def\csname val@rev.desc.d1.sspro.p1.Tsealed.center-d1.tau-source-C.tost1.equivalent\endcsname{0}
\expandafter\def\csname cnt@rev.desc.d1.sspro.p1.Tsealed.center-d1.tau-source-C.tost1.equivalent\endcsname{0}
\expandafter\def\csname pct@rev.desc.d1.sspro.p1.Tsealed.center-d1.tau-source-C.tost1.equivalent\endcsname{0.0}
\expandafter\def\csname ppt@rev.desc.d1.sspro.p1.Tsealed.center-d1.tau-source-C.tost1.equivalent\endcsname{0.0}
\expandafter\def\csname val@rev.desc.d1.sspro.p1.Tsealed.center-d1.tau-source-C.tost1.p\endcsname{0.999}
\expandafter\def\csname cnt@rev.desc.d1.sspro.p1.Tsealed.center-d1.tau-source-C.tost1.p\endcsname{1}
\expandafter\def\csname pct@rev.desc.d1.sspro.p1.Tsealed.center-d1.tau-source-C.tost1.p\endcsname{99.9}
\expandafter\def\csname ppt@rev.desc.d1.sspro.p1.Tsealed.center-d1.tau-source-C.tost1.p\endcsname{$+$99.9}
\expandafter\def\csname val@rev.desc.d1.sspro.p1.Tsealed.center-d1.tau-source-C.tost5.ci90-inside\endcsname{0}
\expandafter\def\csname cnt@rev.desc.d1.sspro.p1.Tsealed.center-d1.tau-source-C.tost5.ci90-inside\endcsname{0}
\expandafter\def\csname pct@rev.desc.d1.sspro.p1.Tsealed.center-d1.tau-source-C.tost5.ci90-inside\endcsname{0.0}
\expandafter\def\csname ppt@rev.desc.d1.sspro.p1.Tsealed.center-d1.tau-source-C.tost5.ci90-inside\endcsname{0.0}
\expandafter\def\csname val@rev.desc.d1.sspro.p1.Tsealed.center-d1.tau-source-C.tost5.equivalent\endcsname{0}
\expandafter\def\csname cnt@rev.desc.d1.sspro.p1.Tsealed.center-d1.tau-source-C.tost5.equivalent\endcsname{0}
\expandafter\def\csname pct@rev.desc.d1.sspro.p1.Tsealed.center-d1.tau-source-C.tost5.equivalent\endcsname{0.0}
\expandafter\def\csname ppt@rev.desc.d1.sspro.p1.Tsealed.center-d1.tau-source-C.tost5.equivalent\endcsname{0.0}
\expandafter\def\csname val@rev.desc.d1.sspro.p1.Tsealed.center-d1.tau-source-C.tost5.p\endcsname{0.938}
\expandafter\def\csname cnt@rev.desc.d1.sspro.p1.Tsealed.center-d1.tau-source-C.tost5.p\endcsname{1}
\expandafter\def\csname pct@rev.desc.d1.sspro.p1.Tsealed.center-d1.tau-source-C.tost5.p\endcsname{93.8}
\expandafter\def\csname ppt@rev.desc.d1.sspro.p1.Tsealed.center-d1.tau-source-C.tost5.p\endcsname{$+$93.8}
\expandafter\def\csname val@rev.desc.d1.sspro.p1.Tsealed.center-d1.tau-target-T.ci90.0\endcsname{0.021}
\expandafter\def\csname cnt@rev.desc.d1.sspro.p1.Tsealed.center-d1.tau-target-T.ci90.0\endcsname{0}
\expandafter\def\csname pct@rev.desc.d1.sspro.p1.Tsealed.center-d1.tau-target-T.ci90.0\endcsname{2.1}
\expandafter\def\csname ppt@rev.desc.d1.sspro.p1.Tsealed.center-d1.tau-target-T.ci90.0\endcsname{$+$2.1}
\expandafter\def\csname val@rev.desc.d1.sspro.p1.Tsealed.center-d1.tau-target-T.ci90.1\endcsname{0.198}
\expandafter\def\csname cnt@rev.desc.d1.sspro.p1.Tsealed.center-d1.tau-target-T.ci90.1\endcsname{0}
\expandafter\def\csname pct@rev.desc.d1.sspro.p1.Tsealed.center-d1.tau-target-T.ci90.1\endcsname{19.8}
\expandafter\def\csname ppt@rev.desc.d1.sspro.p1.Tsealed.center-d1.tau-target-T.ci90.1\endcsname{$+$19.8}
\expandafter\def\csname val@rev.desc.d1.sspro.p1.Tsealed.center-d1.tau-target-T.ci95-excludes-0\endcsname{1}
\expandafter\def\csname cnt@rev.desc.d1.sspro.p1.Tsealed.center-d1.tau-target-T.ci95-excludes-0\endcsname{1}
\expandafter\def\csname pct@rev.desc.d1.sspro.p1.Tsealed.center-d1.tau-target-T.ci95-excludes-0\endcsname{100.0}
\expandafter\def\csname ppt@rev.desc.d1.sspro.p1.Tsealed.center-d1.tau-target-T.ci95-excludes-0\endcsname{$+$100.0}
\expandafter\def\csname val@rev.desc.d1.sspro.p1.Tsealed.center-d1.tau-target-T.ci95.0\endcsname{0.021}
\expandafter\def\csname cnt@rev.desc.d1.sspro.p1.Tsealed.center-d1.tau-target-T.ci95.0\endcsname{0}
\expandafter\def\csname pct@rev.desc.d1.sspro.p1.Tsealed.center-d1.tau-target-T.ci95.0\endcsname{2.1}
\expandafter\def\csname ppt@rev.desc.d1.sspro.p1.Tsealed.center-d1.tau-target-T.ci95.0\endcsname{$+$2.1}
\expandafter\def\csname val@rev.desc.d1.sspro.p1.Tsealed.center-d1.tau-target-T.ci95.1\endcsname{0.219}
\expandafter\def\csname cnt@rev.desc.d1.sspro.p1.Tsealed.center-d1.tau-target-T.ci95.1\endcsname{0}
\expandafter\def\csname pct@rev.desc.d1.sspro.p1.Tsealed.center-d1.tau-target-T.ci95.1\endcsname{21.9}
\expandafter\def\csname ppt@rev.desc.d1.sspro.p1.Tsealed.center-d1.tau-target-T.ci95.1\endcsname{$+$21.9}
\expandafter\def\csname val@rev.desc.d1.sspro.p1.Tsealed.center-d1.tau-target-T.mean\endcsname{0.104}
\expandafter\def\csname cnt@rev.desc.d1.sspro.p1.Tsealed.center-d1.tau-target-T.mean\endcsname{0}
\expandafter\def\csname pct@rev.desc.d1.sspro.p1.Tsealed.center-d1.tau-target-T.mean\endcsname{10.4}
\expandafter\def\csname ppt@rev.desc.d1.sspro.p1.Tsealed.center-d1.tau-target-T.mean\endcsname{$+$10.4}
\expandafter\def\csname val@rev.desc.d1.sspro.p1.Tsealed.center-d1.tau-target-T.n-apps\endcsname{12}
\expandafter\def\csname cnt@rev.desc.d1.sspro.p1.Tsealed.center-d1.tau-target-T.n-apps\endcsname{12}
\expandafter\def\csname pct@rev.desc.d1.sspro.p1.Tsealed.center-d1.tau-target-T.n-apps\endcsname{1200.0}
\expandafter\def\csname ppt@rev.desc.d1.sspro.p1.Tsealed.center-d1.tau-target-T.n-apps\endcsname{$+$1200.0}
\expandafter\def\csname val@rev.desc.d1.sspro.p1.Tsealed.center-d1.tau-target-T.tost1.ci90-inside\endcsname{0}
\expandafter\def\csname cnt@rev.desc.d1.sspro.p1.Tsealed.center-d1.tau-target-T.tost1.ci90-inside\endcsname{0}
\expandafter\def\csname pct@rev.desc.d1.sspro.p1.Tsealed.center-d1.tau-target-T.tost1.ci90-inside\endcsname{0.0}
\expandafter\def\csname ppt@rev.desc.d1.sspro.p1.Tsealed.center-d1.tau-target-T.tost1.ci90-inside\endcsname{0.0}
\expandafter\def\csname val@rev.desc.d1.sspro.p1.Tsealed.center-d1.tau-target-T.tost1.equivalent\endcsname{0}
\expandafter\def\csname cnt@rev.desc.d1.sspro.p1.Tsealed.center-d1.tau-target-T.tost1.equivalent\endcsname{0}
\expandafter\def\csname pct@rev.desc.d1.sspro.p1.Tsealed.center-d1.tau-target-T.tost1.equivalent\endcsname{0.0}
\expandafter\def\csname ppt@rev.desc.d1.sspro.p1.Tsealed.center-d1.tau-target-T.tost1.equivalent\endcsname{0.0}
\expandafter\def\csname val@rev.desc.d1.sspro.p1.Tsealed.center-d1.tau-target-T.tost1.p\endcsname{0.990}
\expandafter\def\csname cnt@rev.desc.d1.sspro.p1.Tsealed.center-d1.tau-target-T.tost1.p\endcsname{1}
\expandafter\def\csname pct@rev.desc.d1.sspro.p1.Tsealed.center-d1.tau-target-T.tost1.p\endcsname{99.0}
\expandafter\def\csname ppt@rev.desc.d1.sspro.p1.Tsealed.center-d1.tau-target-T.tost1.p\endcsname{$+$99.0}
\expandafter\def\csname val@rev.desc.d1.sspro.p1.Tsealed.center-d1.tau-target-T.tost5.ci90-inside\endcsname{0}
\expandafter\def\csname cnt@rev.desc.d1.sspro.p1.Tsealed.center-d1.tau-target-T.tost5.ci90-inside\endcsname{0}
\expandafter\def\csname pct@rev.desc.d1.sspro.p1.Tsealed.center-d1.tau-target-T.tost5.ci90-inside\endcsname{0.0}
\expandafter\def\csname ppt@rev.desc.d1.sspro.p1.Tsealed.center-d1.tau-target-T.tost5.ci90-inside\endcsname{0.0}
\expandafter\def\csname val@rev.desc.d1.sspro.p1.Tsealed.center-d1.tau-target-T.tost5.equivalent\endcsname{0}
\expandafter\def\csname cnt@rev.desc.d1.sspro.p1.Tsealed.center-d1.tau-target-T.tost5.equivalent\endcsname{0}
\expandafter\def\csname pct@rev.desc.d1.sspro.p1.Tsealed.center-d1.tau-target-T.tost5.equivalent\endcsname{0.0}
\expandafter\def\csname ppt@rev.desc.d1.sspro.p1.Tsealed.center-d1.tau-target-T.tost5.equivalent\endcsname{0.0}
\expandafter\def\csname val@rev.desc.d1.sspro.p1.Tsealed.center-d1.tau-target-T.tost5.p\endcsname{0.870}
\expandafter\def\csname cnt@rev.desc.d1.sspro.p1.Tsealed.center-d1.tau-target-T.tost5.p\endcsname{1}
\expandafter\def\csname pct@rev.desc.d1.sspro.p1.Tsealed.center-d1.tau-target-T.tost5.p\endcsname{87.0}
\expandafter\def\csname ppt@rev.desc.d1.sspro.p1.Tsealed.center-d1.tau-target-T.tost5.p\endcsname{$+$87.0}
\expandafter\def\csname val@rev.desc.d1.sspro.p1.Tsealed.center-d1.zero-estimate\endcsname{0}
\expandafter\def\csname cnt@rev.desc.d1.sspro.p1.Tsealed.center-d1.zero-estimate\endcsname{0}
\expandafter\def\csname pct@rev.desc.d1.sspro.p1.Tsealed.center-d1.zero-estimate\endcsname{0.0}
\expandafter\def\csname ppt@rev.desc.d1.sspro.p1.Tsealed.center-d1.zero-estimate\endcsname{0.0}
\expandafter\def\csname val@rev.desc.d1.sspro.p1.Tsealed.context-d1.delta-tau-cross.ci90.0\endcsname{-0.208}
\expandafter\def\csname cnt@rev.desc.d1.sspro.p1.Tsealed.context-d1.delta-tau-cross.ci90.0\endcsname{0}
\expandafter\def\csname pct@rev.desc.d1.sspro.p1.Tsealed.context-d1.delta-tau-cross.ci90.0\endcsname{$-$20.8}
\expandafter\def\csname ppt@rev.desc.d1.sspro.p1.Tsealed.context-d1.delta-tau-cross.ci90.0\endcsname{$-$20.8}
\expandafter\def\csname val@rev.desc.d1.sspro.p1.Tsealed.context-d1.delta-tau-cross.ci90.1\endcsname{0.104}
\expandafter\def\csname cnt@rev.desc.d1.sspro.p1.Tsealed.context-d1.delta-tau-cross.ci90.1\endcsname{0}
\expandafter\def\csname pct@rev.desc.d1.sspro.p1.Tsealed.context-d1.delta-tau-cross.ci90.1\endcsname{10.4}
\expandafter\def\csname ppt@rev.desc.d1.sspro.p1.Tsealed.context-d1.delta-tau-cross.ci90.1\endcsname{$+$10.4}
\expandafter\def\csname val@rev.desc.d1.sspro.p1.Tsealed.context-d1.delta-tau-cross.ci95-excludes-0\endcsname{0}
\expandafter\def\csname cnt@rev.desc.d1.sspro.p1.Tsealed.context-d1.delta-tau-cross.ci95-excludes-0\endcsname{0}
\expandafter\def\csname pct@rev.desc.d1.sspro.p1.Tsealed.context-d1.delta-tau-cross.ci95-excludes-0\endcsname{0.0}
\expandafter\def\csname ppt@rev.desc.d1.sspro.p1.Tsealed.context-d1.delta-tau-cross.ci95-excludes-0\endcsname{0.0}
\expandafter\def\csname val@rev.desc.d1.sspro.p1.Tsealed.context-d1.delta-tau-cross.ci95.0\endcsname{-0.240}
\expandafter\def\csname cnt@rev.desc.d1.sspro.p1.Tsealed.context-d1.delta-tau-cross.ci95.0\endcsname{0}
\expandafter\def\csname pct@rev.desc.d1.sspro.p1.Tsealed.context-d1.delta-tau-cross.ci95.0\endcsname{$-$24.0}
\expandafter\def\csname ppt@rev.desc.d1.sspro.p1.Tsealed.context-d1.delta-tau-cross.ci95.0\endcsname{$-$24.0}
\expandafter\def\csname val@rev.desc.d1.sspro.p1.Tsealed.context-d1.delta-tau-cross.ci95.1\endcsname{0.135}
\expandafter\def\csname cnt@rev.desc.d1.sspro.p1.Tsealed.context-d1.delta-tau-cross.ci95.1\endcsname{0}
\expandafter\def\csname pct@rev.desc.d1.sspro.p1.Tsealed.context-d1.delta-tau-cross.ci95.1\endcsname{13.5}
\expandafter\def\csname ppt@rev.desc.d1.sspro.p1.Tsealed.context-d1.delta-tau-cross.ci95.1\endcsname{$+$13.5}
\expandafter\def\csname val@rev.desc.d1.sspro.p1.Tsealed.context-d1.delta-tau-cross.mean\endcsname{-0.052}
\expandafter\def\csname cnt@rev.desc.d1.sspro.p1.Tsealed.context-d1.delta-tau-cross.mean\endcsname{0}
\expandafter\def\csname pct@rev.desc.d1.sspro.p1.Tsealed.context-d1.delta-tau-cross.mean\endcsname{$-$5.2}
\expandafter\def\csname ppt@rev.desc.d1.sspro.p1.Tsealed.context-d1.delta-tau-cross.mean\endcsname{$-$5.2}
\expandafter\def\csname val@rev.desc.d1.sspro.p1.Tsealed.context-d1.delta-tau-cross.n-apps.0\endcsname{12}
\expandafter\def\csname cnt@rev.desc.d1.sspro.p1.Tsealed.context-d1.delta-tau-cross.n-apps.0\endcsname{12}
\expandafter\def\csname pct@rev.desc.d1.sspro.p1.Tsealed.context-d1.delta-tau-cross.n-apps.0\endcsname{1200.0}
\expandafter\def\csname ppt@rev.desc.d1.sspro.p1.Tsealed.context-d1.delta-tau-cross.n-apps.0\endcsname{$+$1200.0}
\expandafter\def\csname val@rev.desc.d1.sspro.p1.Tsealed.context-d1.delta-tau-cross.n-apps.1\endcsname{6}
\expandafter\def\csname cnt@rev.desc.d1.sspro.p1.Tsealed.context-d1.delta-tau-cross.n-apps.1\endcsname{6}
\expandafter\def\csname pct@rev.desc.d1.sspro.p1.Tsealed.context-d1.delta-tau-cross.n-apps.1\endcsname{600.0}
\expandafter\def\csname ppt@rev.desc.d1.sspro.p1.Tsealed.context-d1.delta-tau-cross.n-apps.1\endcsname{$+$600.0}
\expandafter\def\csname val@rev.desc.d1.sspro.p1.Tsealed.context-d1.delta-tau-cross.tost1.ci90-inside\endcsname{0}
\expandafter\def\csname cnt@rev.desc.d1.sspro.p1.Tsealed.context-d1.delta-tau-cross.tost1.ci90-inside\endcsname{0}
\expandafter\def\csname pct@rev.desc.d1.sspro.p1.Tsealed.context-d1.delta-tau-cross.tost1.ci90-inside\endcsname{0.0}
\expandafter\def\csname ppt@rev.desc.d1.sspro.p1.Tsealed.context-d1.delta-tau-cross.tost1.ci90-inside\endcsname{0.0}
\expandafter\def\csname val@rev.desc.d1.sspro.p1.Tsealed.context-d1.delta-tau-cross.tost1.equivalent\endcsname{0}
\expandafter\def\csname cnt@rev.desc.d1.sspro.p1.Tsealed.context-d1.delta-tau-cross.tost1.equivalent\endcsname{0}
\expandafter\def\csname pct@rev.desc.d1.sspro.p1.Tsealed.context-d1.delta-tau-cross.tost1.equivalent\endcsname{0.0}
\expandafter\def\csname ppt@rev.desc.d1.sspro.p1.Tsealed.context-d1.delta-tau-cross.tost1.equivalent\endcsname{0.0}
\expandafter\def\csname val@rev.desc.d1.sspro.p1.Tsealed.context-d1.delta-tau-cross.tost1.p\endcsname{0.686}
\expandafter\def\csname cnt@rev.desc.d1.sspro.p1.Tsealed.context-d1.delta-tau-cross.tost1.p\endcsname{1}
\expandafter\def\csname pct@rev.desc.d1.sspro.p1.Tsealed.context-d1.delta-tau-cross.tost1.p\endcsname{68.6}
\expandafter\def\csname ppt@rev.desc.d1.sspro.p1.Tsealed.context-d1.delta-tau-cross.tost1.p\endcsname{$+$68.6}
\expandafter\def\csname val@rev.desc.d1.sspro.p1.Tsealed.context-d1.delta-tau-cross.tost5.ci90-inside\endcsname{0}
\expandafter\def\csname cnt@rev.desc.d1.sspro.p1.Tsealed.context-d1.delta-tau-cross.tost5.ci90-inside\endcsname{0}
\expandafter\def\csname pct@rev.desc.d1.sspro.p1.Tsealed.context-d1.delta-tau-cross.tost5.ci90-inside\endcsname{0.0}
\expandafter\def\csname ppt@rev.desc.d1.sspro.p1.Tsealed.context-d1.delta-tau-cross.tost5.ci90-inside\endcsname{0.0}
\expandafter\def\csname val@rev.desc.d1.sspro.p1.Tsealed.context-d1.delta-tau-cross.tost5.equivalent\endcsname{0}
\expandafter\def\csname cnt@rev.desc.d1.sspro.p1.Tsealed.context-d1.delta-tau-cross.tost5.equivalent\endcsname{0}
\expandafter\def\csname pct@rev.desc.d1.sspro.p1.Tsealed.context-d1.delta-tau-cross.tost5.equivalent\endcsname{0.0}
\expandafter\def\csname ppt@rev.desc.d1.sspro.p1.Tsealed.context-d1.delta-tau-cross.tost5.equivalent\endcsname{0.0}
\expandafter\def\csname val@rev.desc.d1.sspro.p1.Tsealed.context-d1.delta-tau-cross.tost5.p\endcsname{0.517}
\expandafter\def\csname cnt@rev.desc.d1.sspro.p1.Tsealed.context-d1.delta-tau-cross.tost5.p\endcsname{1}
\expandafter\def\csname pct@rev.desc.d1.sspro.p1.Tsealed.context-d1.delta-tau-cross.tost5.p\endcsname{51.7}
\expandafter\def\csname ppt@rev.desc.d1.sspro.p1.Tsealed.context-d1.delta-tau-cross.tost5.p\endcsname{$+$51.7}
\expandafter\def\csname val@rev.desc.d1.sspro.p1.Tsealed.context-d1.material-flip\endcsname{0}
\expandafter\def\csname cnt@rev.desc.d1.sspro.p1.Tsealed.context-d1.material-flip\endcsname{0}
\expandafter\def\csname pct@rev.desc.d1.sspro.p1.Tsealed.context-d1.material-flip\endcsname{0.0}
\expandafter\def\csname ppt@rev.desc.d1.sspro.p1.Tsealed.context-d1.material-flip\endcsname{0.0}
\expandafter\def\csname val@rev.desc.d1.sspro.p1.Tsealed.context-d1.n-tasks\endcsname{96}
\expandafter\def\csname cnt@rev.desc.d1.sspro.p1.Tsealed.context-d1.n-tasks\endcsname{96}
\expandafter\def\csname pct@rev.desc.d1.sspro.p1.Tsealed.context-d1.n-tasks\endcsname{9600.0}
\expandafter\def\csname ppt@rev.desc.d1.sspro.p1.Tsealed.context-d1.n-tasks\endcsname{$+$9600.0}
\expandafter\def\csname val@rev.desc.d1.sspro.p1.Tsealed.context-d1.n-tasks-source-C\endcsname{48}
\expandafter\def\csname cnt@rev.desc.d1.sspro.p1.Tsealed.context-d1.n-tasks-source-C\endcsname{48}
\expandafter\def\csname pct@rev.desc.d1.sspro.p1.Tsealed.context-d1.n-tasks-source-C\endcsname{4800.0}
\expandafter\def\csname ppt@rev.desc.d1.sspro.p1.Tsealed.context-d1.n-tasks-source-C\endcsname{$+$4800.0}
\expandafter\def\csname val@rev.desc.d1.sspro.p1.Tsealed.context-d1.paper-tau-T\endcsname{0.115}
\expandafter\def\csname cnt@rev.desc.d1.sspro.p1.Tsealed.context-d1.paper-tau-T\endcsname{0}
\expandafter\def\csname pct@rev.desc.d1.sspro.p1.Tsealed.context-d1.paper-tau-T\endcsname{11.5}
\expandafter\def\csname ppt@rev.desc.d1.sspro.p1.Tsealed.context-d1.paper-tau-T\endcsname{$+$11.5}
\expandafter\def\csname val@rev.desc.d1.sspro.p1.Tsealed.context-d1.reproduces\endcsname{1}
\expandafter\def\csname cnt@rev.desc.d1.sspro.p1.Tsealed.context-d1.reproduces\endcsname{1}
\expandafter\def\csname pct@rev.desc.d1.sspro.p1.Tsealed.context-d1.reproduces\endcsname{100.0}
\expandafter\def\csname ppt@rev.desc.d1.sspro.p1.Tsealed.context-d1.reproduces\endcsname{$+$100.0}
\expandafter\def\csname val@rev.desc.d1.sspro.p1.Tsealed.context-d1.reproduces-ci\endcsname{1}
\expandafter\def\csname cnt@rev.desc.d1.sspro.p1.Tsealed.context-d1.reproduces-ci\endcsname{1}
\expandafter\def\csname pct@rev.desc.d1.sspro.p1.Tsealed.context-d1.reproduces-ci\endcsname{100.0}
\expandafter\def\csname ppt@rev.desc.d1.sspro.p1.Tsealed.context-d1.reproduces-ci\endcsname{$+$100.0}
\expandafter\def\csname val@rev.desc.d1.sspro.p1.Tsealed.context-d1.sign-differs\endcsname{0}
\expandafter\def\csname cnt@rev.desc.d1.sspro.p1.Tsealed.context-d1.sign-differs\endcsname{0}
\expandafter\def\csname pct@rev.desc.d1.sspro.p1.Tsealed.context-d1.sign-differs\endcsname{0.0}
\expandafter\def\csname ppt@rev.desc.d1.sspro.p1.Tsealed.context-d1.sign-differs\endcsname{0.0}
\expandafter\def\csname val@rev.desc.d1.sspro.p1.Tsealed.context-d1.tau-source-C.ci90.0\endcsname{0.062}
\expandafter\def\csname cnt@rev.desc.d1.sspro.p1.Tsealed.context-d1.tau-source-C.ci90.0\endcsname{0}
\expandafter\def\csname pct@rev.desc.d1.sspro.p1.Tsealed.context-d1.tau-source-C.ci90.0\endcsname{6.2}
\expandafter\def\csname ppt@rev.desc.d1.sspro.p1.Tsealed.context-d1.tau-source-C.ci90.0\endcsname{$+$6.2}
\expandafter\def\csname val@rev.desc.d1.sspro.p1.Tsealed.context-d1.tau-source-C.ci90.1\endcsname{0.292}
\expandafter\def\csname cnt@rev.desc.d1.sspro.p1.Tsealed.context-d1.tau-source-C.ci90.1\endcsname{0}
\expandafter\def\csname pct@rev.desc.d1.sspro.p1.Tsealed.context-d1.tau-source-C.ci90.1\endcsname{29.2}
\expandafter\def\csname ppt@rev.desc.d1.sspro.p1.Tsealed.context-d1.tau-source-C.ci90.1\endcsname{$+$29.2}
\expandafter\def\csname val@rev.desc.d1.sspro.p1.Tsealed.context-d1.tau-source-C.ci95-excludes-0\endcsname{1}
\expandafter\def\csname cnt@rev.desc.d1.sspro.p1.Tsealed.context-d1.tau-source-C.ci95-excludes-0\endcsname{1}
\expandafter\def\csname pct@rev.desc.d1.sspro.p1.Tsealed.context-d1.tau-source-C.ci95-excludes-0\endcsname{100.0}
\expandafter\def\csname ppt@rev.desc.d1.sspro.p1.Tsealed.context-d1.tau-source-C.ci95-excludes-0\endcsname{$+$100.0}
\expandafter\def\csname val@rev.desc.d1.sspro.p1.Tsealed.context-d1.tau-source-C.ci95.0\endcsname{0.042}
\expandafter\def\csname cnt@rev.desc.d1.sspro.p1.Tsealed.context-d1.tau-source-C.ci95.0\endcsname{0}
\expandafter\def\csname pct@rev.desc.d1.sspro.p1.Tsealed.context-d1.tau-source-C.ci95.0\endcsname{4.2}
\expandafter\def\csname ppt@rev.desc.d1.sspro.p1.Tsealed.context-d1.tau-source-C.ci95.0\endcsname{$+$4.2}
\expandafter\def\csname val@rev.desc.d1.sspro.p1.Tsealed.context-d1.tau-source-C.ci95.1\endcsname{0.312}
\expandafter\def\csname cnt@rev.desc.d1.sspro.p1.Tsealed.context-d1.tau-source-C.ci95.1\endcsname{0}
\expandafter\def\csname pct@rev.desc.d1.sspro.p1.Tsealed.context-d1.tau-source-C.ci95.1\endcsname{31.2}
\expandafter\def\csname ppt@rev.desc.d1.sspro.p1.Tsealed.context-d1.tau-source-C.ci95.1\endcsname{$+$31.2}
\expandafter\def\csname val@rev.desc.d1.sspro.p1.Tsealed.context-d1.tau-source-C.mean\endcsname{0.167}
\expandafter\def\csname cnt@rev.desc.d1.sspro.p1.Tsealed.context-d1.tau-source-C.mean\endcsname{0}
\expandafter\def\csname pct@rev.desc.d1.sspro.p1.Tsealed.context-d1.tau-source-C.mean\endcsname{16.7}
\expandafter\def\csname ppt@rev.desc.d1.sspro.p1.Tsealed.context-d1.tau-source-C.mean\endcsname{$+$16.7}
\expandafter\def\csname val@rev.desc.d1.sspro.p1.Tsealed.context-d1.tau-source-C.n-apps\endcsname{6}
\expandafter\def\csname cnt@rev.desc.d1.sspro.p1.Tsealed.context-d1.tau-source-C.n-apps\endcsname{6}
\expandafter\def\csname pct@rev.desc.d1.sspro.p1.Tsealed.context-d1.tau-source-C.n-apps\endcsname{600.0}
\expandafter\def\csname ppt@rev.desc.d1.sspro.p1.Tsealed.context-d1.tau-source-C.n-apps\endcsname{$+$600.0}
\expandafter\def\csname val@rev.desc.d1.sspro.p1.Tsealed.context-d1.tau-source-C.tost1.ci90-inside\endcsname{0}
\expandafter\def\csname cnt@rev.desc.d1.sspro.p1.Tsealed.context-d1.tau-source-C.tost1.ci90-inside\endcsname{0}
\expandafter\def\csname pct@rev.desc.d1.sspro.p1.Tsealed.context-d1.tau-source-C.tost1.ci90-inside\endcsname{0.0}
\expandafter\def\csname ppt@rev.desc.d1.sspro.p1.Tsealed.context-d1.tau-source-C.tost1.ci90-inside\endcsname{0.0}
\expandafter\def\csname val@rev.desc.d1.sspro.p1.Tsealed.context-d1.tau-source-C.tost1.equivalent\endcsname{0}
\expandafter\def\csname cnt@rev.desc.d1.sspro.p1.Tsealed.context-d1.tau-source-C.tost1.equivalent\endcsname{0}
\expandafter\def\csname pct@rev.desc.d1.sspro.p1.Tsealed.context-d1.tau-source-C.tost1.equivalent\endcsname{0.0}
\expandafter\def\csname ppt@rev.desc.d1.sspro.p1.Tsealed.context-d1.tau-source-C.tost1.equivalent\endcsname{0.0}
\expandafter\def\csname val@rev.desc.d1.sspro.p1.Tsealed.context-d1.tau-source-C.tost1.p\endcsname{0.997}
\expandafter\def\csname cnt@rev.desc.d1.sspro.p1.Tsealed.context-d1.tau-source-C.tost1.p\endcsname{1}
\expandafter\def\csname pct@rev.desc.d1.sspro.p1.Tsealed.context-d1.tau-source-C.tost1.p\endcsname{99.7}
\expandafter\def\csname ppt@rev.desc.d1.sspro.p1.Tsealed.context-d1.tau-source-C.tost1.p\endcsname{$+$99.7}
\expandafter\def\csname val@rev.desc.d1.sspro.p1.Tsealed.context-d1.tau-source-C.tost5.ci90-inside\endcsname{0}
\expandafter\def\csname cnt@rev.desc.d1.sspro.p1.Tsealed.context-d1.tau-source-C.tost5.ci90-inside\endcsname{0}
\expandafter\def\csname pct@rev.desc.d1.sspro.p1.Tsealed.context-d1.tau-source-C.tost5.ci90-inside\endcsname{0.0}
\expandafter\def\csname ppt@rev.desc.d1.sspro.p1.Tsealed.context-d1.tau-source-C.tost5.ci90-inside\endcsname{0.0}
\expandafter\def\csname val@rev.desc.d1.sspro.p1.Tsealed.context-d1.tau-source-C.tost5.equivalent\endcsname{0}
\expandafter\def\csname cnt@rev.desc.d1.sspro.p1.Tsealed.context-d1.tau-source-C.tost5.equivalent\endcsname{0}
\expandafter\def\csname pct@rev.desc.d1.sspro.p1.Tsealed.context-d1.tau-source-C.tost5.equivalent\endcsname{0.0}
\expandafter\def\csname ppt@rev.desc.d1.sspro.p1.Tsealed.context-d1.tau-source-C.tost5.equivalent\endcsname{0.0}
\expandafter\def\csname val@rev.desc.d1.sspro.p1.Tsealed.context-d1.tau-source-C.tost5.p\endcsname{0.968}
\expandafter\def\csname cnt@rev.desc.d1.sspro.p1.Tsealed.context-d1.tau-source-C.tost5.p\endcsname{1}
\expandafter\def\csname pct@rev.desc.d1.sspro.p1.Tsealed.context-d1.tau-source-C.tost5.p\endcsname{96.8}
\expandafter\def\csname ppt@rev.desc.d1.sspro.p1.Tsealed.context-d1.tau-source-C.tost5.p\endcsname{$+$96.8}
\expandafter\def\csname val@rev.desc.d1.sspro.p1.Tsealed.context-d1.tau-target-T.ci90.0\endcsname{0.031}
\expandafter\def\csname cnt@rev.desc.d1.sspro.p1.Tsealed.context-d1.tau-target-T.ci90.0\endcsname{0}
\expandafter\def\csname pct@rev.desc.d1.sspro.p1.Tsealed.context-d1.tau-target-T.ci90.0\endcsname{3.1}
\expandafter\def\csname ppt@rev.desc.d1.sspro.p1.Tsealed.context-d1.tau-target-T.ci90.0\endcsname{$+$3.1}
\expandafter\def\csname val@rev.desc.d1.sspro.p1.Tsealed.context-d1.tau-target-T.ci90.1\endcsname{0.229}
\expandafter\def\csname cnt@rev.desc.d1.sspro.p1.Tsealed.context-d1.tau-target-T.ci90.1\endcsname{0}
\expandafter\def\csname pct@rev.desc.d1.sspro.p1.Tsealed.context-d1.tau-target-T.ci90.1\endcsname{22.9}
\expandafter\def\csname ppt@rev.desc.d1.sspro.p1.Tsealed.context-d1.tau-target-T.ci90.1\endcsname{$+$22.9}
\expandafter\def\csname val@rev.desc.d1.sspro.p1.Tsealed.context-d1.tau-target-T.ci95-excludes-0\endcsname{1}
\expandafter\def\csname cnt@rev.desc.d1.sspro.p1.Tsealed.context-d1.tau-target-T.ci95-excludes-0\endcsname{1}
\expandafter\def\csname pct@rev.desc.d1.sspro.p1.Tsealed.context-d1.tau-target-T.ci95-excludes-0\endcsname{100.0}
\expandafter\def\csname ppt@rev.desc.d1.sspro.p1.Tsealed.context-d1.tau-target-T.ci95-excludes-0\endcsname{$+$100.0}
\expandafter\def\csname val@rev.desc.d1.sspro.p1.Tsealed.context-d1.tau-target-T.ci95.0\endcsname{0.021}
\expandafter\def\csname cnt@rev.desc.d1.sspro.p1.Tsealed.context-d1.tau-target-T.ci95.0\endcsname{0}
\expandafter\def\csname pct@rev.desc.d1.sspro.p1.Tsealed.context-d1.tau-target-T.ci95.0\endcsname{2.1}
\expandafter\def\csname ppt@rev.desc.d1.sspro.p1.Tsealed.context-d1.tau-target-T.ci95.0\endcsname{$+$2.1}
\expandafter\def\csname val@rev.desc.d1.sspro.p1.Tsealed.context-d1.tau-target-T.ci95.1\endcsname{0.260}
\expandafter\def\csname cnt@rev.desc.d1.sspro.p1.Tsealed.context-d1.tau-target-T.ci95.1\endcsname{0}
\expandafter\def\csname pct@rev.desc.d1.sspro.p1.Tsealed.context-d1.tau-target-T.ci95.1\endcsname{26.0}
\expandafter\def\csname ppt@rev.desc.d1.sspro.p1.Tsealed.context-d1.tau-target-T.ci95.1\endcsname{$+$26.0}
\expandafter\def\csname val@rev.desc.d1.sspro.p1.Tsealed.context-d1.tau-target-T.mean\endcsname{0.115}
\expandafter\def\csname cnt@rev.desc.d1.sspro.p1.Tsealed.context-d1.tau-target-T.mean\endcsname{0}
\expandafter\def\csname pct@rev.desc.d1.sspro.p1.Tsealed.context-d1.tau-target-T.mean\endcsname{11.5}
\expandafter\def\csname ppt@rev.desc.d1.sspro.p1.Tsealed.context-d1.tau-target-T.mean\endcsname{$+$11.5}
\expandafter\def\csname val@rev.desc.d1.sspro.p1.Tsealed.context-d1.tau-target-T.n-apps\endcsname{12}
\expandafter\def\csname cnt@rev.desc.d1.sspro.p1.Tsealed.context-d1.tau-target-T.n-apps\endcsname{12}
\expandafter\def\csname pct@rev.desc.d1.sspro.p1.Tsealed.context-d1.tau-target-T.n-apps\endcsname{1200.0}
\expandafter\def\csname ppt@rev.desc.d1.sspro.p1.Tsealed.context-d1.tau-target-T.n-apps\endcsname{$+$1200.0}
\expandafter\def\csname val@rev.desc.d1.sspro.p1.Tsealed.context-d1.tau-target-T.tost1.ci90-inside\endcsname{0}
\expandafter\def\csname cnt@rev.desc.d1.sspro.p1.Tsealed.context-d1.tau-target-T.tost1.ci90-inside\endcsname{0}
\expandafter\def\csname pct@rev.desc.d1.sspro.p1.Tsealed.context-d1.tau-target-T.tost1.ci90-inside\endcsname{0.0}
\expandafter\def\csname ppt@rev.desc.d1.sspro.p1.Tsealed.context-d1.tau-target-T.tost1.ci90-inside\endcsname{0.0}
\expandafter\def\csname val@rev.desc.d1.sspro.p1.Tsealed.context-d1.tau-target-T.tost1.equivalent\endcsname{0}
\expandafter\def\csname cnt@rev.desc.d1.sspro.p1.Tsealed.context-d1.tau-target-T.tost1.equivalent\endcsname{0}
\expandafter\def\csname pct@rev.desc.d1.sspro.p1.Tsealed.context-d1.tau-target-T.tost1.equivalent\endcsname{0.0}
\expandafter\def\csname ppt@rev.desc.d1.sspro.p1.Tsealed.context-d1.tau-target-T.tost1.equivalent\endcsname{0.0}
\expandafter\def\csname val@rev.desc.d1.sspro.p1.Tsealed.context-d1.tau-target-T.tost1.p\endcsname{0.989}
\expandafter\def\csname cnt@rev.desc.d1.sspro.p1.Tsealed.context-d1.tau-target-T.tost1.p\endcsname{1}
\expandafter\def\csname pct@rev.desc.d1.sspro.p1.Tsealed.context-d1.tau-target-T.tost1.p\endcsname{98.9}
\expandafter\def\csname ppt@rev.desc.d1.sspro.p1.Tsealed.context-d1.tau-target-T.tost1.p\endcsname{$+$98.9}
\expandafter\def\csname val@rev.desc.d1.sspro.p1.Tsealed.context-d1.tau-target-T.tost5.ci90-inside\endcsname{0}
\expandafter\def\csname cnt@rev.desc.d1.sspro.p1.Tsealed.context-d1.tau-target-T.tost5.ci90-inside\endcsname{0}
\expandafter\def\csname pct@rev.desc.d1.sspro.p1.Tsealed.context-d1.tau-target-T.tost5.ci90-inside\endcsname{0.0}
\expandafter\def\csname ppt@rev.desc.d1.sspro.p1.Tsealed.context-d1.tau-target-T.tost5.ci90-inside\endcsname{0.0}
\expandafter\def\csname val@rev.desc.d1.sspro.p1.Tsealed.context-d1.tau-target-T.tost5.equivalent\endcsname{0}
\expandafter\def\csname cnt@rev.desc.d1.sspro.p1.Tsealed.context-d1.tau-target-T.tost5.equivalent\endcsname{0}
\expandafter\def\csname pct@rev.desc.d1.sspro.p1.Tsealed.context-d1.tau-target-T.tost5.equivalent\endcsname{0.0}
\expandafter\def\csname ppt@rev.desc.d1.sspro.p1.Tsealed.context-d1.tau-target-T.tost5.equivalent\endcsname{0.0}
\expandafter\def\csname val@rev.desc.d1.sspro.p1.Tsealed.context-d1.tau-target-T.tost5.p\endcsname{0.865}
\expandafter\def\csname cnt@rev.desc.d1.sspro.p1.Tsealed.context-d1.tau-target-T.tost5.p\endcsname{1}
\expandafter\def\csname pct@rev.desc.d1.sspro.p1.Tsealed.context-d1.tau-target-T.tost5.p\endcsname{86.5}
\expandafter\def\csname ppt@rev.desc.d1.sspro.p1.Tsealed.context-d1.tau-target-T.tost5.p\endcsname{$+$86.5}
\expandafter\def\csname val@rev.desc.d1.sspro.p1.Tsealed.context-d1.zero-estimate\endcsname{0}
\expandafter\def\csname cnt@rev.desc.d1.sspro.p1.Tsealed.context-d1.zero-estimate\endcsname{0}
\expandafter\def\csname pct@rev.desc.d1.sspro.p1.Tsealed.context-d1.zero-estimate\endcsname{0.0}
\expandafter\def\csname ppt@rev.desc.d1.sspro.p1.Tsealed.context-d1.zero-estimate\endcsname{0.0}
\expandafter\def\csname val@rev.desc.d1.sspro.p1.Tsealed.dual-d2.delta-tau-cross.ci90.0\endcsname{-0.167}
\expandafter\def\csname cnt@rev.desc.d1.sspro.p1.Tsealed.dual-d2.delta-tau-cross.ci90.0\endcsname{0}
\expandafter\def\csname pct@rev.desc.d1.sspro.p1.Tsealed.dual-d2.delta-tau-cross.ci90.0\endcsname{$-$16.7}
\expandafter\def\csname ppt@rev.desc.d1.sspro.p1.Tsealed.dual-d2.delta-tau-cross.ci90.0\endcsname{$-$16.7}
\expandafter\def\csname val@rev.desc.d1.sspro.p1.Tsealed.dual-d2.delta-tau-cross.ci90.1\endcsname{0.208}
\expandafter\def\csname cnt@rev.desc.d1.sspro.p1.Tsealed.dual-d2.delta-tau-cross.ci90.1\endcsname{0}
\expandafter\def\csname pct@rev.desc.d1.sspro.p1.Tsealed.dual-d2.delta-tau-cross.ci90.1\endcsname{20.8}
\expandafter\def\csname ppt@rev.desc.d1.sspro.p1.Tsealed.dual-d2.delta-tau-cross.ci90.1\endcsname{$+$20.8}
\expandafter\def\csname val@rev.desc.d1.sspro.p1.Tsealed.dual-d2.delta-tau-cross.ci95-excludes-0\endcsname{0}
\expandafter\def\csname cnt@rev.desc.d1.sspro.p1.Tsealed.dual-d2.delta-tau-cross.ci95-excludes-0\endcsname{0}
\expandafter\def\csname pct@rev.desc.d1.sspro.p1.Tsealed.dual-d2.delta-tau-cross.ci95-excludes-0\endcsname{0.0}
\expandafter\def\csname ppt@rev.desc.d1.sspro.p1.Tsealed.dual-d2.delta-tau-cross.ci95-excludes-0\endcsname{0.0}
\expandafter\def\csname val@rev.desc.d1.sspro.p1.Tsealed.dual-d2.delta-tau-cross.ci95.0\endcsname{-0.198}
\expandafter\def\csname cnt@rev.desc.d1.sspro.p1.Tsealed.dual-d2.delta-tau-cross.ci95.0\endcsname{0}
\expandafter\def\csname pct@rev.desc.d1.sspro.p1.Tsealed.dual-d2.delta-tau-cross.ci95.0\endcsname{$-$19.8}
\expandafter\def\csname ppt@rev.desc.d1.sspro.p1.Tsealed.dual-d2.delta-tau-cross.ci95.0\endcsname{$-$19.8}
\expandafter\def\csname val@rev.desc.d1.sspro.p1.Tsealed.dual-d2.delta-tau-cross.ci95.1\endcsname{0.250}
\expandafter\def\csname cnt@rev.desc.d1.sspro.p1.Tsealed.dual-d2.delta-tau-cross.ci95.1\endcsname{0}
\expandafter\def\csname pct@rev.desc.d1.sspro.p1.Tsealed.dual-d2.delta-tau-cross.ci95.1\endcsname{25.0}
\expandafter\def\csname ppt@rev.desc.d1.sspro.p1.Tsealed.dual-d2.delta-tau-cross.ci95.1\endcsname{$+$25.0}
\expandafter\def\csname val@rev.desc.d1.sspro.p1.Tsealed.dual-d2.delta-tau-cross.mean\endcsname{0.021}
\expandafter\def\csname cnt@rev.desc.d1.sspro.p1.Tsealed.dual-d2.delta-tau-cross.mean\endcsname{0}
\expandafter\def\csname pct@rev.desc.d1.sspro.p1.Tsealed.dual-d2.delta-tau-cross.mean\endcsname{2.1}
\expandafter\def\csname ppt@rev.desc.d1.sspro.p1.Tsealed.dual-d2.delta-tau-cross.mean\endcsname{$+$2.1}
\expandafter\def\csname val@rev.desc.d1.sspro.p1.Tsealed.dual-d2.delta-tau-cross.n-apps.0\endcsname{12}
\expandafter\def\csname cnt@rev.desc.d1.sspro.p1.Tsealed.dual-d2.delta-tau-cross.n-apps.0\endcsname{12}
\expandafter\def\csname pct@rev.desc.d1.sspro.p1.Tsealed.dual-d2.delta-tau-cross.n-apps.0\endcsname{1200.0}
\expandafter\def\csname ppt@rev.desc.d1.sspro.p1.Tsealed.dual-d2.delta-tau-cross.n-apps.0\endcsname{$+$1200.0}
\expandafter\def\csname val@rev.desc.d1.sspro.p1.Tsealed.dual-d2.delta-tau-cross.n-apps.1\endcsname{6}
\expandafter\def\csname cnt@rev.desc.d1.sspro.p1.Tsealed.dual-d2.delta-tau-cross.n-apps.1\endcsname{6}
\expandafter\def\csname pct@rev.desc.d1.sspro.p1.Tsealed.dual-d2.delta-tau-cross.n-apps.1\endcsname{600.0}
\expandafter\def\csname ppt@rev.desc.d1.sspro.p1.Tsealed.dual-d2.delta-tau-cross.n-apps.1\endcsname{$+$600.0}
\expandafter\def\csname val@rev.desc.d1.sspro.p1.Tsealed.dual-d2.delta-tau-cross.tost1.ci90-inside\endcsname{0}
\expandafter\def\csname cnt@rev.desc.d1.sspro.p1.Tsealed.dual-d2.delta-tau-cross.tost1.ci90-inside\endcsname{0}
\expandafter\def\csname pct@rev.desc.d1.sspro.p1.Tsealed.dual-d2.delta-tau-cross.tost1.ci90-inside\endcsname{0.0}
\expandafter\def\csname ppt@rev.desc.d1.sspro.p1.Tsealed.dual-d2.delta-tau-cross.tost1.ci90-inside\endcsname{0.0}
\expandafter\def\csname val@rev.desc.d1.sspro.p1.Tsealed.dual-d2.delta-tau-cross.tost1.equivalent\endcsname{0}
\expandafter\def\csname cnt@rev.desc.d1.sspro.p1.Tsealed.dual-d2.delta-tau-cross.tost1.equivalent\endcsname{0}
\expandafter\def\csname pct@rev.desc.d1.sspro.p1.Tsealed.dual-d2.delta-tau-cross.tost1.equivalent\endcsname{0.0}
\expandafter\def\csname ppt@rev.desc.d1.sspro.p1.Tsealed.dual-d2.delta-tau-cross.tost1.equivalent\endcsname{0.0}
\expandafter\def\csname val@rev.desc.d1.sspro.p1.Tsealed.dual-d2.delta-tau-cross.tost1.p\endcsname{0.553}
\expandafter\def\csname cnt@rev.desc.d1.sspro.p1.Tsealed.dual-d2.delta-tau-cross.tost1.p\endcsname{1}
\expandafter\def\csname pct@rev.desc.d1.sspro.p1.Tsealed.dual-d2.delta-tau-cross.tost1.p\endcsname{55.3}
\expandafter\def\csname ppt@rev.desc.d1.sspro.p1.Tsealed.dual-d2.delta-tau-cross.tost1.p\endcsname{$+$55.3}
\expandafter\def\csname val@rev.desc.d1.sspro.p1.Tsealed.dual-d2.delta-tau-cross.tost5.ci90-inside\endcsname{0}
\expandafter\def\csname cnt@rev.desc.d1.sspro.p1.Tsealed.dual-d2.delta-tau-cross.tost5.ci90-inside\endcsname{0}
\expandafter\def\csname pct@rev.desc.d1.sspro.p1.Tsealed.dual-d2.delta-tau-cross.tost5.ci90-inside\endcsname{0.0}
\expandafter\def\csname ppt@rev.desc.d1.sspro.p1.Tsealed.dual-d2.delta-tau-cross.tost5.ci90-inside\endcsname{0.0}
\expandafter\def\csname val@rev.desc.d1.sspro.p1.Tsealed.dual-d2.delta-tau-cross.tost5.equivalent\endcsname{0}
\expandafter\def\csname cnt@rev.desc.d1.sspro.p1.Tsealed.dual-d2.delta-tau-cross.tost5.equivalent\endcsname{0}
\expandafter\def\csname pct@rev.desc.d1.sspro.p1.Tsealed.dual-d2.delta-tau-cross.tost5.equivalent\endcsname{0.0}
\expandafter\def\csname ppt@rev.desc.d1.sspro.p1.Tsealed.dual-d2.delta-tau-cross.tost5.equivalent\endcsname{0.0}
\expandafter\def\csname val@rev.desc.d1.sspro.p1.Tsealed.dual-d2.delta-tau-cross.tost5.p\endcsname{0.401}
\expandafter\def\csname cnt@rev.desc.d1.sspro.p1.Tsealed.dual-d2.delta-tau-cross.tost5.p\endcsname{0}
\expandafter\def\csname pct@rev.desc.d1.sspro.p1.Tsealed.dual-d2.delta-tau-cross.tost5.p\endcsname{40.1}
\expandafter\def\csname ppt@rev.desc.d1.sspro.p1.Tsealed.dual-d2.delta-tau-cross.tost5.p\endcsname{$+$40.1}
\expandafter\def\csname val@rev.desc.d1.sspro.p1.Tsealed.dual-d2.material-flip\endcsname{0}
\expandafter\def\csname cnt@rev.desc.d1.sspro.p1.Tsealed.dual-d2.material-flip\endcsname{0}
\expandafter\def\csname pct@rev.desc.d1.sspro.p1.Tsealed.dual-d2.material-flip\endcsname{0.0}
\expandafter\def\csname ppt@rev.desc.d1.sspro.p1.Tsealed.dual-d2.material-flip\endcsname{0.0}
\expandafter\def\csname val@rev.desc.d1.sspro.p1.Tsealed.dual-d2.n-tasks\endcsname{96}
\expandafter\def\csname cnt@rev.desc.d1.sspro.p1.Tsealed.dual-d2.n-tasks\endcsname{96}
\expandafter\def\csname pct@rev.desc.d1.sspro.p1.Tsealed.dual-d2.n-tasks\endcsname{9600.0}
\expandafter\def\csname ppt@rev.desc.d1.sspro.p1.Tsealed.dual-d2.n-tasks\endcsname{$+$9600.0}
\expandafter\def\csname val@rev.desc.d1.sspro.p1.Tsealed.dual-d2.n-tasks-source-C\endcsname{48}
\expandafter\def\csname cnt@rev.desc.d1.sspro.p1.Tsealed.dual-d2.n-tasks-source-C\endcsname{48}
\expandafter\def\csname pct@rev.desc.d1.sspro.p1.Tsealed.dual-d2.n-tasks-source-C\endcsname{4800.0}
\expandafter\def\csname ppt@rev.desc.d1.sspro.p1.Tsealed.dual-d2.n-tasks-source-C\endcsname{$+$4800.0}
\expandafter\def\csname val@rev.desc.d1.sspro.p1.Tsealed.dual-d2.paper-tau-T\endcsname{0.146}
\expandafter\def\csname cnt@rev.desc.d1.sspro.p1.Tsealed.dual-d2.paper-tau-T\endcsname{0}
\expandafter\def\csname pct@rev.desc.d1.sspro.p1.Tsealed.dual-d2.paper-tau-T\endcsname{14.6}
\expandafter\def\csname ppt@rev.desc.d1.sspro.p1.Tsealed.dual-d2.paper-tau-T\endcsname{$+$14.6}
\expandafter\def\csname val@rev.desc.d1.sspro.p1.Tsealed.dual-d2.reproduces\endcsname{1}
\expandafter\def\csname cnt@rev.desc.d1.sspro.p1.Tsealed.dual-d2.reproduces\endcsname{1}
\expandafter\def\csname pct@rev.desc.d1.sspro.p1.Tsealed.dual-d2.reproduces\endcsname{100.0}
\expandafter\def\csname ppt@rev.desc.d1.sspro.p1.Tsealed.dual-d2.reproduces\endcsname{$+$100.0}
\expandafter\def\csname val@rev.desc.d1.sspro.p1.Tsealed.dual-d2.reproduces-ci\endcsname{1}
\expandafter\def\csname cnt@rev.desc.d1.sspro.p1.Tsealed.dual-d2.reproduces-ci\endcsname{1}
\expandafter\def\csname pct@rev.desc.d1.sspro.p1.Tsealed.dual-d2.reproduces-ci\endcsname{100.0}
\expandafter\def\csname ppt@rev.desc.d1.sspro.p1.Tsealed.dual-d2.reproduces-ci\endcsname{$+$100.0}
\expandafter\def\csname val@rev.desc.d1.sspro.p1.Tsealed.dual-d2.sign-differs\endcsname{0}
\expandafter\def\csname cnt@rev.desc.d1.sspro.p1.Tsealed.dual-d2.sign-differs\endcsname{0}
\expandafter\def\csname pct@rev.desc.d1.sspro.p1.Tsealed.dual-d2.sign-differs\endcsname{0.0}
\expandafter\def\csname ppt@rev.desc.d1.sspro.p1.Tsealed.dual-d2.sign-differs\endcsname{0.0}
\expandafter\def\csname val@rev.desc.d1.sspro.p1.Tsealed.dual-d2.tau-source-C.ci90.0\endcsname{-0.021}
\expandafter\def\csname cnt@rev.desc.d1.sspro.p1.Tsealed.dual-d2.tau-source-C.ci90.0\endcsname{0}
\expandafter\def\csname pct@rev.desc.d1.sspro.p1.Tsealed.dual-d2.tau-source-C.ci90.0\endcsname{$-$2.1}
\expandafter\def\csname ppt@rev.desc.d1.sspro.p1.Tsealed.dual-d2.tau-source-C.ci90.0\endcsname{$-$2.1}
\expandafter\def\csname val@rev.desc.d1.sspro.p1.Tsealed.dual-d2.tau-source-C.ci90.1\endcsname{0.271}
\expandafter\def\csname cnt@rev.desc.d1.sspro.p1.Tsealed.dual-d2.tau-source-C.ci90.1\endcsname{0}
\expandafter\def\csname pct@rev.desc.d1.sspro.p1.Tsealed.dual-d2.tau-source-C.ci90.1\endcsname{27.1}
\expandafter\def\csname ppt@rev.desc.d1.sspro.p1.Tsealed.dual-d2.tau-source-C.ci90.1\endcsname{$+$27.1}
\expandafter\def\csname val@rev.desc.d1.sspro.p1.Tsealed.dual-d2.tau-source-C.ci95-excludes-0\endcsname{0}
\expandafter\def\csname cnt@rev.desc.d1.sspro.p1.Tsealed.dual-d2.tau-source-C.ci95-excludes-0\endcsname{0}
\expandafter\def\csname pct@rev.desc.d1.sspro.p1.Tsealed.dual-d2.tau-source-C.ci95-excludes-0\endcsname{0.0}
\expandafter\def\csname ppt@rev.desc.d1.sspro.p1.Tsealed.dual-d2.tau-source-C.ci95-excludes-0\endcsname{0.0}
\expandafter\def\csname val@rev.desc.d1.sspro.p1.Tsealed.dual-d2.tau-source-C.ci95.0\endcsname{-0.042}
\expandafter\def\csname cnt@rev.desc.d1.sspro.p1.Tsealed.dual-d2.tau-source-C.ci95.0\endcsname{0}
\expandafter\def\csname pct@rev.desc.d1.sspro.p1.Tsealed.dual-d2.tau-source-C.ci95.0\endcsname{$-$4.2}
\expandafter\def\csname ppt@rev.desc.d1.sspro.p1.Tsealed.dual-d2.tau-source-C.ci95.0\endcsname{$-$4.2}
\expandafter\def\csname val@rev.desc.d1.sspro.p1.Tsealed.dual-d2.tau-source-C.ci95.1\endcsname{0.292}
\expandafter\def\csname cnt@rev.desc.d1.sspro.p1.Tsealed.dual-d2.tau-source-C.ci95.1\endcsname{0}
\expandafter\def\csname pct@rev.desc.d1.sspro.p1.Tsealed.dual-d2.tau-source-C.ci95.1\endcsname{29.2}
\expandafter\def\csname ppt@rev.desc.d1.sspro.p1.Tsealed.dual-d2.tau-source-C.ci95.1\endcsname{$+$29.2}
\expandafter\def\csname val@rev.desc.d1.sspro.p1.Tsealed.dual-d2.tau-source-C.mean\endcsname{0.125}
\expandafter\def\csname cnt@rev.desc.d1.sspro.p1.Tsealed.dual-d2.tau-source-C.mean\endcsname{0}
\expandafter\def\csname pct@rev.desc.d1.sspro.p1.Tsealed.dual-d2.tau-source-C.mean\endcsname{12.5}
\expandafter\def\csname ppt@rev.desc.d1.sspro.p1.Tsealed.dual-d2.tau-source-C.mean\endcsname{$+$12.5}
\expandafter\def\csname val@rev.desc.d1.sspro.p1.Tsealed.dual-d2.tau-source-C.n-apps\endcsname{6}
\expandafter\def\csname cnt@rev.desc.d1.sspro.p1.Tsealed.dual-d2.tau-source-C.n-apps\endcsname{6}
\expandafter\def\csname pct@rev.desc.d1.sspro.p1.Tsealed.dual-d2.tau-source-C.n-apps\endcsname{600.0}
\expandafter\def\csname ppt@rev.desc.d1.sspro.p1.Tsealed.dual-d2.tau-source-C.n-apps\endcsname{$+$600.0}
\expandafter\def\csname val@rev.desc.d1.sspro.p1.Tsealed.dual-d2.tau-source-C.tost1.ci90-inside\endcsname{0}
\expandafter\def\csname cnt@rev.desc.d1.sspro.p1.Tsealed.dual-d2.tau-source-C.tost1.ci90-inside\endcsname{0}
\expandafter\def\csname pct@rev.desc.d1.sspro.p1.Tsealed.dual-d2.tau-source-C.tost1.ci90-inside\endcsname{0.0}
\expandafter\def\csname ppt@rev.desc.d1.sspro.p1.Tsealed.dual-d2.tau-source-C.tost1.ci90-inside\endcsname{0.0}
\expandafter\def\csname val@rev.desc.d1.sspro.p1.Tsealed.dual-d2.tau-source-C.tost1.equivalent\endcsname{0}
\expandafter\def\csname cnt@rev.desc.d1.sspro.p1.Tsealed.dual-d2.tau-source-C.tost1.equivalent\endcsname{0}
\expandafter\def\csname pct@rev.desc.d1.sspro.p1.Tsealed.dual-d2.tau-source-C.tost1.equivalent\endcsname{0.0}
\expandafter\def\csname ppt@rev.desc.d1.sspro.p1.Tsealed.dual-d2.tau-source-C.tost1.equivalent\endcsname{0.0}
\expandafter\def\csname val@rev.desc.d1.sspro.p1.Tsealed.dual-d2.tau-source-C.tost1.p\endcsname{0.904}
\expandafter\def\csname cnt@rev.desc.d1.sspro.p1.Tsealed.dual-d2.tau-source-C.tost1.p\endcsname{1}
\expandafter\def\csname pct@rev.desc.d1.sspro.p1.Tsealed.dual-d2.tau-source-C.tost1.p\endcsname{90.4}
\expandafter\def\csname ppt@rev.desc.d1.sspro.p1.Tsealed.dual-d2.tau-source-C.tost1.p\endcsname{$+$90.4}
\expandafter\def\csname val@rev.desc.d1.sspro.p1.Tsealed.dual-d2.tau-source-C.tost5.ci90-inside\endcsname{0}
\expandafter\def\csname cnt@rev.desc.d1.sspro.p1.Tsealed.dual-d2.tau-source-C.tost5.ci90-inside\endcsname{0}
\expandafter\def\csname pct@rev.desc.d1.sspro.p1.Tsealed.dual-d2.tau-source-C.tost5.ci90-inside\endcsname{0.0}
\expandafter\def\csname ppt@rev.desc.d1.sspro.p1.Tsealed.dual-d2.tau-source-C.tost5.ci90-inside\endcsname{0.0}
\expandafter\def\csname val@rev.desc.d1.sspro.p1.Tsealed.dual-d2.tau-source-C.tost5.equivalent\endcsname{0}
\expandafter\def\csname cnt@rev.desc.d1.sspro.p1.Tsealed.dual-d2.tau-source-C.tost5.equivalent\endcsname{0}
\expandafter\def\csname pct@rev.desc.d1.sspro.p1.Tsealed.dual-d2.tau-source-C.tost5.equivalent\endcsname{0.0}
\expandafter\def\csname ppt@rev.desc.d1.sspro.p1.Tsealed.dual-d2.tau-source-C.tost5.equivalent\endcsname{0.0}
\expandafter\def\csname val@rev.desc.d1.sspro.p1.Tsealed.dual-d2.tau-source-C.tost5.p\endcsname{0.806}
\expandafter\def\csname cnt@rev.desc.d1.sspro.p1.Tsealed.dual-d2.tau-source-C.tost5.p\endcsname{1}
\expandafter\def\csname pct@rev.desc.d1.sspro.p1.Tsealed.dual-d2.tau-source-C.tost5.p\endcsname{80.6}
\expandafter\def\csname ppt@rev.desc.d1.sspro.p1.Tsealed.dual-d2.tau-source-C.tost5.p\endcsname{$+$80.6}
\expandafter\def\csname val@rev.desc.d1.sspro.p1.Tsealed.dual-d2.tau-target-T.ci90.0\endcsname{0.031}
\expandafter\def\csname cnt@rev.desc.d1.sspro.p1.Tsealed.dual-d2.tau-target-T.ci90.0\endcsname{0}
\expandafter\def\csname pct@rev.desc.d1.sspro.p1.Tsealed.dual-d2.tau-target-T.ci90.0\endcsname{3.1}
\expandafter\def\csname ppt@rev.desc.d1.sspro.p1.Tsealed.dual-d2.tau-target-T.ci90.0\endcsname{$+$3.1}
\expandafter\def\csname val@rev.desc.d1.sspro.p1.Tsealed.dual-d2.tau-target-T.ci90.1\endcsname{0.271}
\expandafter\def\csname cnt@rev.desc.d1.sspro.p1.Tsealed.dual-d2.tau-target-T.ci90.1\endcsname{0}
\expandafter\def\csname pct@rev.desc.d1.sspro.p1.Tsealed.dual-d2.tau-target-T.ci90.1\endcsname{27.1}
\expandafter\def\csname ppt@rev.desc.d1.sspro.p1.Tsealed.dual-d2.tau-target-T.ci90.1\endcsname{$+$27.1}
\expandafter\def\csname val@rev.desc.d1.sspro.p1.Tsealed.dual-d2.tau-target-T.ci95-excludes-0\endcsname{1}
\expandafter\def\csname cnt@rev.desc.d1.sspro.p1.Tsealed.dual-d2.tau-target-T.ci95-excludes-0\endcsname{1}
\expandafter\def\csname pct@rev.desc.d1.sspro.p1.Tsealed.dual-d2.tau-target-T.ci95-excludes-0\endcsname{100.0}
\expandafter\def\csname ppt@rev.desc.d1.sspro.p1.Tsealed.dual-d2.tau-target-T.ci95-excludes-0\endcsname{$+$100.0}
\expandafter\def\csname val@rev.desc.d1.sspro.p1.Tsealed.dual-d2.tau-target-T.ci95.0\endcsname{0.010}
\expandafter\def\csname cnt@rev.desc.d1.sspro.p1.Tsealed.dual-d2.tau-target-T.ci95.0\endcsname{0}
\expandafter\def\csname pct@rev.desc.d1.sspro.p1.Tsealed.dual-d2.tau-target-T.ci95.0\endcsname{1.0}
\expandafter\def\csname ppt@rev.desc.d1.sspro.p1.Tsealed.dual-d2.tau-target-T.ci95.0\endcsname{$+$1.0}
\expandafter\def\csname val@rev.desc.d1.sspro.p1.Tsealed.dual-d2.tau-target-T.ci95.1\endcsname{0.302}
\expandafter\def\csname cnt@rev.desc.d1.sspro.p1.Tsealed.dual-d2.tau-target-T.ci95.1\endcsname{0}
\expandafter\def\csname pct@rev.desc.d1.sspro.p1.Tsealed.dual-d2.tau-target-T.ci95.1\endcsname{30.2}
\expandafter\def\csname ppt@rev.desc.d1.sspro.p1.Tsealed.dual-d2.tau-target-T.ci95.1\endcsname{$+$30.2}
\expandafter\def\csname val@rev.desc.d1.sspro.p1.Tsealed.dual-d2.tau-target-T.mean\endcsname{0.146}
\expandafter\def\csname cnt@rev.desc.d1.sspro.p1.Tsealed.dual-d2.tau-target-T.mean\endcsname{0}
\expandafter\def\csname pct@rev.desc.d1.sspro.p1.Tsealed.dual-d2.tau-target-T.mean\endcsname{14.6}
\expandafter\def\csname ppt@rev.desc.d1.sspro.p1.Tsealed.dual-d2.tau-target-T.mean\endcsname{$+$14.6}
\expandafter\def\csname val@rev.desc.d1.sspro.p1.Tsealed.dual-d2.tau-target-T.n-apps\endcsname{12}
\expandafter\def\csname cnt@rev.desc.d1.sspro.p1.Tsealed.dual-d2.tau-target-T.n-apps\endcsname{12}
\expandafter\def\csname pct@rev.desc.d1.sspro.p1.Tsealed.dual-d2.tau-target-T.n-apps\endcsname{1200.0}
\expandafter\def\csname ppt@rev.desc.d1.sspro.p1.Tsealed.dual-d2.tau-target-T.n-apps\endcsname{$+$1200.0}
\expandafter\def\csname val@rev.desc.d1.sspro.p1.Tsealed.dual-d2.tau-target-T.tost1.ci90-inside\endcsname{0}
\expandafter\def\csname cnt@rev.desc.d1.sspro.p1.Tsealed.dual-d2.tau-target-T.tost1.ci90-inside\endcsname{0}
\expandafter\def\csname pct@rev.desc.d1.sspro.p1.Tsealed.dual-d2.tau-target-T.tost1.ci90-inside\endcsname{0.0}
\expandafter\def\csname ppt@rev.desc.d1.sspro.p1.Tsealed.dual-d2.tau-target-T.tost1.ci90-inside\endcsname{0.0}
\expandafter\def\csname val@rev.desc.d1.sspro.p1.Tsealed.dual-d2.tau-target-T.tost1.equivalent\endcsname{0}
\expandafter\def\csname cnt@rev.desc.d1.sspro.p1.Tsealed.dual-d2.tau-target-T.tost1.equivalent\endcsname{0}
\expandafter\def\csname pct@rev.desc.d1.sspro.p1.Tsealed.dual-d2.tau-target-T.tost1.equivalent\endcsname{0.0}
\expandafter\def\csname ppt@rev.desc.d1.sspro.p1.Tsealed.dual-d2.tau-target-T.tost1.equivalent\endcsname{0.0}
\expandafter\def\csname val@rev.desc.d1.sspro.p1.Tsealed.dual-d2.tau-target-T.tost1.p\endcsname{0.980}
\expandafter\def\csname cnt@rev.desc.d1.sspro.p1.Tsealed.dual-d2.tau-target-T.tost1.p\endcsname{1}
\expandafter\def\csname pct@rev.desc.d1.sspro.p1.Tsealed.dual-d2.tau-target-T.tost1.p\endcsname{98.0}
\expandafter\def\csname ppt@rev.desc.d1.sspro.p1.Tsealed.dual-d2.tau-target-T.tost1.p\endcsname{$+$98.0}
\expandafter\def\csname val@rev.desc.d1.sspro.p1.Tsealed.dual-d2.tau-target-T.tost5.ci90-inside\endcsname{0}
\expandafter\def\csname cnt@rev.desc.d1.sspro.p1.Tsealed.dual-d2.tau-target-T.tost5.ci90-inside\endcsname{0}
\expandafter\def\csname pct@rev.desc.d1.sspro.p1.Tsealed.dual-d2.tau-target-T.tost5.ci90-inside\endcsname{0.0}
\expandafter\def\csname ppt@rev.desc.d1.sspro.p1.Tsealed.dual-d2.tau-target-T.tost5.ci90-inside\endcsname{0.0}
\expandafter\def\csname val@rev.desc.d1.sspro.p1.Tsealed.dual-d2.tau-target-T.tost5.equivalent\endcsname{0}
\expandafter\def\csname cnt@rev.desc.d1.sspro.p1.Tsealed.dual-d2.tau-target-T.tost5.equivalent\endcsname{0}
\expandafter\def\csname pct@rev.desc.d1.sspro.p1.Tsealed.dual-d2.tau-target-T.tost5.equivalent\endcsname{0.0}
\expandafter\def\csname ppt@rev.desc.d1.sspro.p1.Tsealed.dual-d2.tau-target-T.tost5.equivalent\endcsname{0.0}
\expandafter\def\csname val@rev.desc.d1.sspro.p1.Tsealed.dual-d2.tau-target-T.tost5.p\endcsname{0.920}
\expandafter\def\csname cnt@rev.desc.d1.sspro.p1.Tsealed.dual-d2.tau-target-T.tost5.p\endcsname{1}
\expandafter\def\csname pct@rev.desc.d1.sspro.p1.Tsealed.dual-d2.tau-target-T.tost5.p\endcsname{92.0}
\expandafter\def\csname ppt@rev.desc.d1.sspro.p1.Tsealed.dual-d2.tau-target-T.tost5.p\endcsname{$+$92.0}
\expandafter\def\csname val@rev.desc.d1.sspro.p1.Tsealed.dual-d2.zero-estimate\endcsname{0}
\expandafter\def\csname cnt@rev.desc.d1.sspro.p1.Tsealed.dual-d2.zero-estimate\endcsname{0}
\expandafter\def\csname pct@rev.desc.d1.sspro.p1.Tsealed.dual-d2.zero-estimate\endcsname{0.0}
\expandafter\def\csname ppt@rev.desc.d1.sspro.p1.Tsealed.dual-d2.zero-estimate\endcsname{0.0}
\expandafter\def\csname val@rev.desc.d1.sspro.p1.Tsealed.grid2-d3.delta-tau-cross.ci90.0\endcsname{-0.250}
\expandafter\def\csname cnt@rev.desc.d1.sspro.p1.Tsealed.grid2-d3.delta-tau-cross.ci90.0\endcsname{0}
\expandafter\def\csname pct@rev.desc.d1.sspro.p1.Tsealed.grid2-d3.delta-tau-cross.ci90.0\endcsname{$-$25.0}
\expandafter\def\csname ppt@rev.desc.d1.sspro.p1.Tsealed.grid2-d3.delta-tau-cross.ci90.0\endcsname{$-$25.0}
\expandafter\def\csname val@rev.desc.d1.sspro.p1.Tsealed.grid2-d3.delta-tau-cross.ci90.1\endcsname{0.063}
\expandafter\def\csname cnt@rev.desc.d1.sspro.p1.Tsealed.grid2-d3.delta-tau-cross.ci90.1\endcsname{0}
\expandafter\def\csname pct@rev.desc.d1.sspro.p1.Tsealed.grid2-d3.delta-tau-cross.ci90.1\endcsname{6.3}
\expandafter\def\csname ppt@rev.desc.d1.sspro.p1.Tsealed.grid2-d3.delta-tau-cross.ci90.1\endcsname{$+$6.3}
\expandafter\def\csname val@rev.desc.d1.sspro.p1.Tsealed.grid2-d3.delta-tau-cross.ci95-excludes-0\endcsname{0}
\expandafter\def\csname cnt@rev.desc.d1.sspro.p1.Tsealed.grid2-d3.delta-tau-cross.ci95-excludes-0\endcsname{0}
\expandafter\def\csname pct@rev.desc.d1.sspro.p1.Tsealed.grid2-d3.delta-tau-cross.ci95-excludes-0\endcsname{0.0}
\expandafter\def\csname ppt@rev.desc.d1.sspro.p1.Tsealed.grid2-d3.delta-tau-cross.ci95-excludes-0\endcsname{0.0}
\expandafter\def\csname val@rev.desc.d1.sspro.p1.Tsealed.grid2-d3.delta-tau-cross.ci95.0\endcsname{-0.281}
\expandafter\def\csname cnt@rev.desc.d1.sspro.p1.Tsealed.grid2-d3.delta-tau-cross.ci95.0\endcsname{0}
\expandafter\def\csname pct@rev.desc.d1.sspro.p1.Tsealed.grid2-d3.delta-tau-cross.ci95.0\endcsname{$-$28.1}
\expandafter\def\csname ppt@rev.desc.d1.sspro.p1.Tsealed.grid2-d3.delta-tau-cross.ci95.0\endcsname{$-$28.1}
\expandafter\def\csname val@rev.desc.d1.sspro.p1.Tsealed.grid2-d3.delta-tau-cross.ci95.1\endcsname{0.104}
\expandafter\def\csname cnt@rev.desc.d1.sspro.p1.Tsealed.grid2-d3.delta-tau-cross.ci95.1\endcsname{0}
\expandafter\def\csname pct@rev.desc.d1.sspro.p1.Tsealed.grid2-d3.delta-tau-cross.ci95.1\endcsname{10.4}
\expandafter\def\csname ppt@rev.desc.d1.sspro.p1.Tsealed.grid2-d3.delta-tau-cross.ci95.1\endcsname{$+$10.4}
\expandafter\def\csname val@rev.desc.d1.sspro.p1.Tsealed.grid2-d3.delta-tau-cross.mean\endcsname{-0.094}
\expandafter\def\csname cnt@rev.desc.d1.sspro.p1.Tsealed.grid2-d3.delta-tau-cross.mean\endcsname{0}
\expandafter\def\csname pct@rev.desc.d1.sspro.p1.Tsealed.grid2-d3.delta-tau-cross.mean\endcsname{$-$9.4}
\expandafter\def\csname ppt@rev.desc.d1.sspro.p1.Tsealed.grid2-d3.delta-tau-cross.mean\endcsname{$-$9.4}
\expandafter\def\csname val@rev.desc.d1.sspro.p1.Tsealed.grid2-d3.delta-tau-cross.n-apps.0\endcsname{12}
\expandafter\def\csname cnt@rev.desc.d1.sspro.p1.Tsealed.grid2-d3.delta-tau-cross.n-apps.0\endcsname{12}
\expandafter\def\csname pct@rev.desc.d1.sspro.p1.Tsealed.grid2-d3.delta-tau-cross.n-apps.0\endcsname{1200.0}
\expandafter\def\csname ppt@rev.desc.d1.sspro.p1.Tsealed.grid2-d3.delta-tau-cross.n-apps.0\endcsname{$+$1200.0}
\expandafter\def\csname val@rev.desc.d1.sspro.p1.Tsealed.grid2-d3.delta-tau-cross.n-apps.1\endcsname{6}
\expandafter\def\csname cnt@rev.desc.d1.sspro.p1.Tsealed.grid2-d3.delta-tau-cross.n-apps.1\endcsname{6}
\expandafter\def\csname pct@rev.desc.d1.sspro.p1.Tsealed.grid2-d3.delta-tau-cross.n-apps.1\endcsname{600.0}
\expandafter\def\csname ppt@rev.desc.d1.sspro.p1.Tsealed.grid2-d3.delta-tau-cross.n-apps.1\endcsname{$+$600.0}
\expandafter\def\csname val@rev.desc.d1.sspro.p1.Tsealed.grid2-d3.delta-tau-cross.tost1.ci90-inside\endcsname{0}
\expandafter\def\csname cnt@rev.desc.d1.sspro.p1.Tsealed.grid2-d3.delta-tau-cross.tost1.ci90-inside\endcsname{0}
\expandafter\def\csname pct@rev.desc.d1.sspro.p1.Tsealed.grid2-d3.delta-tau-cross.tost1.ci90-inside\endcsname{0.0}
\expandafter\def\csname ppt@rev.desc.d1.sspro.p1.Tsealed.grid2-d3.delta-tau-cross.tost1.ci90-inside\endcsname{0.0}
\expandafter\def\csname val@rev.desc.d1.sspro.p1.Tsealed.grid2-d3.delta-tau-cross.tost1.equivalent\endcsname{0}
\expandafter\def\csname cnt@rev.desc.d1.sspro.p1.Tsealed.grid2-d3.delta-tau-cross.tost1.equivalent\endcsname{0}
\expandafter\def\csname pct@rev.desc.d1.sspro.p1.Tsealed.grid2-d3.delta-tau-cross.tost1.equivalent\endcsname{0.0}
\expandafter\def\csname ppt@rev.desc.d1.sspro.p1.Tsealed.grid2-d3.delta-tau-cross.tost1.equivalent\endcsname{0.0}
\expandafter\def\csname val@rev.desc.d1.sspro.p1.Tsealed.grid2-d3.delta-tau-cross.tost1.p\endcsname{0.817}
\expandafter\def\csname cnt@rev.desc.d1.sspro.p1.Tsealed.grid2-d3.delta-tau-cross.tost1.p\endcsname{1}
\expandafter\def\csname pct@rev.desc.d1.sspro.p1.Tsealed.grid2-d3.delta-tau-cross.tost1.p\endcsname{81.7}
\expandafter\def\csname ppt@rev.desc.d1.sspro.p1.Tsealed.grid2-d3.delta-tau-cross.tost1.p\endcsname{$+$81.7}
\expandafter\def\csname val@rev.desc.d1.sspro.p1.Tsealed.grid2-d3.delta-tau-cross.tost5.ci90-inside\endcsname{0}
\expandafter\def\csname cnt@rev.desc.d1.sspro.p1.Tsealed.grid2-d3.delta-tau-cross.tost5.ci90-inside\endcsname{0}
\expandafter\def\csname pct@rev.desc.d1.sspro.p1.Tsealed.grid2-d3.delta-tau-cross.tost5.ci90-inside\endcsname{0.0}
\expandafter\def\csname ppt@rev.desc.d1.sspro.p1.Tsealed.grid2-d3.delta-tau-cross.tost5.ci90-inside\endcsname{0.0}
\expandafter\def\csname val@rev.desc.d1.sspro.p1.Tsealed.grid2-d3.delta-tau-cross.tost5.equivalent\endcsname{0}
\expandafter\def\csname cnt@rev.desc.d1.sspro.p1.Tsealed.grid2-d3.delta-tau-cross.tost5.equivalent\endcsname{0}
\expandafter\def\csname pct@rev.desc.d1.sspro.p1.Tsealed.grid2-d3.delta-tau-cross.tost5.equivalent\endcsname{0.0}
\expandafter\def\csname ppt@rev.desc.d1.sspro.p1.Tsealed.grid2-d3.delta-tau-cross.tost5.equivalent\endcsname{0.0}
\expandafter\def\csname val@rev.desc.d1.sspro.p1.Tsealed.grid2-d3.delta-tau-cross.tost5.p\endcsname{0.686}
\expandafter\def\csname cnt@rev.desc.d1.sspro.p1.Tsealed.grid2-d3.delta-tau-cross.tost5.p\endcsname{1}
\expandafter\def\csname pct@rev.desc.d1.sspro.p1.Tsealed.grid2-d3.delta-tau-cross.tost5.p\endcsname{68.6}
\expandafter\def\csname ppt@rev.desc.d1.sspro.p1.Tsealed.grid2-d3.delta-tau-cross.tost5.p\endcsname{$+$68.6}
\expandafter\def\csname val@rev.desc.d1.sspro.p1.Tsealed.grid2-d3.material-flip\endcsname{0}
\expandafter\def\csname cnt@rev.desc.d1.sspro.p1.Tsealed.grid2-d3.material-flip\endcsname{0}
\expandafter\def\csname pct@rev.desc.d1.sspro.p1.Tsealed.grid2-d3.material-flip\endcsname{0.0}
\expandafter\def\csname ppt@rev.desc.d1.sspro.p1.Tsealed.grid2-d3.material-flip\endcsname{0.0}
\expandafter\def\csname val@rev.desc.d1.sspro.p1.Tsealed.grid2-d3.n-tasks\endcsname{96}
\expandafter\def\csname cnt@rev.desc.d1.sspro.p1.Tsealed.grid2-d3.n-tasks\endcsname{96}
\expandafter\def\csname pct@rev.desc.d1.sspro.p1.Tsealed.grid2-d3.n-tasks\endcsname{9600.0}
\expandafter\def\csname ppt@rev.desc.d1.sspro.p1.Tsealed.grid2-d3.n-tasks\endcsname{$+$9600.0}
\expandafter\def\csname val@rev.desc.d1.sspro.p1.Tsealed.grid2-d3.n-tasks-source-C\endcsname{48}
\expandafter\def\csname cnt@rev.desc.d1.sspro.p1.Tsealed.grid2-d3.n-tasks-source-C\endcsname{48}
\expandafter\def\csname pct@rev.desc.d1.sspro.p1.Tsealed.grid2-d3.n-tasks-source-C\endcsname{4800.0}
\expandafter\def\csname ppt@rev.desc.d1.sspro.p1.Tsealed.grid2-d3.n-tasks-source-C\endcsname{$+$4800.0}
\expandafter\def\csname val@rev.desc.d1.sspro.p1.Tsealed.grid2-d3.paper-tau-T\endcsname{0.115}
\expandafter\def\csname cnt@rev.desc.d1.sspro.p1.Tsealed.grid2-d3.paper-tau-T\endcsname{0}
\expandafter\def\csname pct@rev.desc.d1.sspro.p1.Tsealed.grid2-d3.paper-tau-T\endcsname{11.5}
\expandafter\def\csname ppt@rev.desc.d1.sspro.p1.Tsealed.grid2-d3.paper-tau-T\endcsname{$+$11.5}
\expandafter\def\csname val@rev.desc.d1.sspro.p1.Tsealed.grid2-d3.reproduces\endcsname{1}
\expandafter\def\csname cnt@rev.desc.d1.sspro.p1.Tsealed.grid2-d3.reproduces\endcsname{1}
\expandafter\def\csname pct@rev.desc.d1.sspro.p1.Tsealed.grid2-d3.reproduces\endcsname{100.0}
\expandafter\def\csname ppt@rev.desc.d1.sspro.p1.Tsealed.grid2-d3.reproduces\endcsname{$+$100.0}
\expandafter\def\csname val@rev.desc.d1.sspro.p1.Tsealed.grid2-d3.reproduces-ci\endcsname{1}
\expandafter\def\csname cnt@rev.desc.d1.sspro.p1.Tsealed.grid2-d3.reproduces-ci\endcsname{1}
\expandafter\def\csname pct@rev.desc.d1.sspro.p1.Tsealed.grid2-d3.reproduces-ci\endcsname{100.0}
\expandafter\def\csname ppt@rev.desc.d1.sspro.p1.Tsealed.grid2-d3.reproduces-ci\endcsname{$+$100.0}
\expandafter\def\csname val@rev.desc.d1.sspro.p1.Tsealed.grid2-d3.sign-differs\endcsname{0}
\expandafter\def\csname cnt@rev.desc.d1.sspro.p1.Tsealed.grid2-d3.sign-differs\endcsname{0}
\expandafter\def\csname pct@rev.desc.d1.sspro.p1.Tsealed.grid2-d3.sign-differs\endcsname{0.0}
\expandafter\def\csname ppt@rev.desc.d1.sspro.p1.Tsealed.grid2-d3.sign-differs\endcsname{0.0}
\expandafter\def\csname val@rev.desc.d1.sspro.p1.Tsealed.grid2-d3.tau-source-C.ci90.0\endcsname{0.083}
\expandafter\def\csname cnt@rev.desc.d1.sspro.p1.Tsealed.grid2-d3.tau-source-C.ci90.0\endcsname{0}
\expandafter\def\csname pct@rev.desc.d1.sspro.p1.Tsealed.grid2-d3.tau-source-C.ci90.0\endcsname{8.3}
\expandafter\def\csname ppt@rev.desc.d1.sspro.p1.Tsealed.grid2-d3.tau-source-C.ci90.0\endcsname{$+$8.3}
\expandafter\def\csname val@rev.desc.d1.sspro.p1.Tsealed.grid2-d3.tau-source-C.ci90.1\endcsname{0.333}
\expandafter\def\csname cnt@rev.desc.d1.sspro.p1.Tsealed.grid2-d3.tau-source-C.ci90.1\endcsname{0}
\expandafter\def\csname pct@rev.desc.d1.sspro.p1.Tsealed.grid2-d3.tau-source-C.ci90.1\endcsname{33.3}
\expandafter\def\csname ppt@rev.desc.d1.sspro.p1.Tsealed.grid2-d3.tau-source-C.ci90.1\endcsname{$+$33.3}
\expandafter\def\csname val@rev.desc.d1.sspro.p1.Tsealed.grid2-d3.tau-source-C.ci95-excludes-0\endcsname{1}
\expandafter\def\csname cnt@rev.desc.d1.sspro.p1.Tsealed.grid2-d3.tau-source-C.ci95-excludes-0\endcsname{1}
\expandafter\def\csname pct@rev.desc.d1.sspro.p1.Tsealed.grid2-d3.tau-source-C.ci95-excludes-0\endcsname{100.0}
\expandafter\def\csname ppt@rev.desc.d1.sspro.p1.Tsealed.grid2-d3.tau-source-C.ci95-excludes-0\endcsname{$+$100.0}
\expandafter\def\csname val@rev.desc.d1.sspro.p1.Tsealed.grid2-d3.tau-source-C.ci95.0\endcsname{0.062}
\expandafter\def\csname cnt@rev.desc.d1.sspro.p1.Tsealed.grid2-d3.tau-source-C.ci95.0\endcsname{0}
\expandafter\def\csname pct@rev.desc.d1.sspro.p1.Tsealed.grid2-d3.tau-source-C.ci95.0\endcsname{6.2}
\expandafter\def\csname ppt@rev.desc.d1.sspro.p1.Tsealed.grid2-d3.tau-source-C.ci95.0\endcsname{$+$6.2}
\expandafter\def\csname val@rev.desc.d1.sspro.p1.Tsealed.grid2-d3.tau-source-C.ci95.1\endcsname{0.354}
\expandafter\def\csname cnt@rev.desc.d1.sspro.p1.Tsealed.grid2-d3.tau-source-C.ci95.1\endcsname{0}
\expandafter\def\csname pct@rev.desc.d1.sspro.p1.Tsealed.grid2-d3.tau-source-C.ci95.1\endcsname{35.4}
\expandafter\def\csname ppt@rev.desc.d1.sspro.p1.Tsealed.grid2-d3.tau-source-C.ci95.1\endcsname{$+$35.4}
\expandafter\def\csname val@rev.desc.d1.sspro.p1.Tsealed.grid2-d3.tau-source-C.mean\endcsname{0.208}
\expandafter\def\csname cnt@rev.desc.d1.sspro.p1.Tsealed.grid2-d3.tau-source-C.mean\endcsname{0}
\expandafter\def\csname pct@rev.desc.d1.sspro.p1.Tsealed.grid2-d3.tau-source-C.mean\endcsname{20.8}
\expandafter\def\csname ppt@rev.desc.d1.sspro.p1.Tsealed.grid2-d3.tau-source-C.mean\endcsname{$+$20.8}
\expandafter\def\csname val@rev.desc.d1.sspro.p1.Tsealed.grid2-d3.tau-source-C.n-apps\endcsname{6}
\expandafter\def\csname cnt@rev.desc.d1.sspro.p1.Tsealed.grid2-d3.tau-source-C.n-apps\endcsname{6}
\expandafter\def\csname pct@rev.desc.d1.sspro.p1.Tsealed.grid2-d3.tau-source-C.n-apps\endcsname{600.0}
\expandafter\def\csname ppt@rev.desc.d1.sspro.p1.Tsealed.grid2-d3.tau-source-C.n-apps\endcsname{$+$600.0}
\expandafter\def\csname val@rev.desc.d1.sspro.p1.Tsealed.grid2-d3.tau-source-C.tost1.ci90-inside\endcsname{0}
\expandafter\def\csname cnt@rev.desc.d1.sspro.p1.Tsealed.grid2-d3.tau-source-C.tost1.ci90-inside\endcsname{0}
\expandafter\def\csname pct@rev.desc.d1.sspro.p1.Tsealed.grid2-d3.tau-source-C.tost1.ci90-inside\endcsname{0.0}
\expandafter\def\csname ppt@rev.desc.d1.sspro.p1.Tsealed.grid2-d3.tau-source-C.tost1.ci90-inside\endcsname{0.0}
\expandafter\def\csname val@rev.desc.d1.sspro.p1.Tsealed.grid2-d3.tau-source-C.tost1.equivalent\endcsname{0}
\expandafter\def\csname cnt@rev.desc.d1.sspro.p1.Tsealed.grid2-d3.tau-source-C.tost1.equivalent\endcsname{0}
\expandafter\def\csname pct@rev.desc.d1.sspro.p1.Tsealed.grid2-d3.tau-source-C.tost1.equivalent\endcsname{0.0}
\expandafter\def\csname ppt@rev.desc.d1.sspro.p1.Tsealed.grid2-d3.tau-source-C.tost1.equivalent\endcsname{0.0}
\expandafter\def\csname val@rev.desc.d1.sspro.p1.Tsealed.grid2-d3.tau-source-C.tost1.p\endcsname{0.999}
\expandafter\def\csname cnt@rev.desc.d1.sspro.p1.Tsealed.grid2-d3.tau-source-C.tost1.p\endcsname{1}
\expandafter\def\csname pct@rev.desc.d1.sspro.p1.Tsealed.grid2-d3.tau-source-C.tost1.p\endcsname{99.9}
\expandafter\def\csname ppt@rev.desc.d1.sspro.p1.Tsealed.grid2-d3.tau-source-C.tost1.p\endcsname{$+$99.9}
\expandafter\def\csname val@rev.desc.d1.sspro.p1.Tsealed.grid2-d3.tau-source-C.tost5.ci90-inside\endcsname{0}
\expandafter\def\csname cnt@rev.desc.d1.sspro.p1.Tsealed.grid2-d3.tau-source-C.tost5.ci90-inside\endcsname{0}
\expandafter\def\csname pct@rev.desc.d1.sspro.p1.Tsealed.grid2-d3.tau-source-C.tost5.ci90-inside\endcsname{0.0}
\expandafter\def\csname ppt@rev.desc.d1.sspro.p1.Tsealed.grid2-d3.tau-source-C.tost5.ci90-inside\endcsname{0.0}
\expandafter\def\csname val@rev.desc.d1.sspro.p1.Tsealed.grid2-d3.tau-source-C.tost5.equivalent\endcsname{0}
\expandafter\def\csname cnt@rev.desc.d1.sspro.p1.Tsealed.grid2-d3.tau-source-C.tost5.equivalent\endcsname{0}
\expandafter\def\csname pct@rev.desc.d1.sspro.p1.Tsealed.grid2-d3.tau-source-C.tost5.equivalent\endcsname{0.0}
\expandafter\def\csname ppt@rev.desc.d1.sspro.p1.Tsealed.grid2-d3.tau-source-C.tost5.equivalent\endcsname{0.0}
\expandafter\def\csname val@rev.desc.d1.sspro.p1.Tsealed.grid2-d3.tau-source-C.tost5.p\endcsname{0.987}
\expandafter\def\csname cnt@rev.desc.d1.sspro.p1.Tsealed.grid2-d3.tau-source-C.tost5.p\endcsname{1}
\expandafter\def\csname pct@rev.desc.d1.sspro.p1.Tsealed.grid2-d3.tau-source-C.tost5.p\endcsname{98.7}
\expandafter\def\csname ppt@rev.desc.d1.sspro.p1.Tsealed.grid2-d3.tau-source-C.tost5.p\endcsname{$+$98.7}
\expandafter\def\csname val@rev.desc.d1.sspro.p1.Tsealed.grid2-d3.tau-target-T.ci90.0\endcsname{0.021}
\expandafter\def\csname cnt@rev.desc.d1.sspro.p1.Tsealed.grid2-d3.tau-target-T.ci90.0\endcsname{0}
\expandafter\def\csname pct@rev.desc.d1.sspro.p1.Tsealed.grid2-d3.tau-target-T.ci90.0\endcsname{2.1}
\expandafter\def\csname ppt@rev.desc.d1.sspro.p1.Tsealed.grid2-d3.tau-target-T.ci90.0\endcsname{$+$2.1}
\expandafter\def\csname val@rev.desc.d1.sspro.p1.Tsealed.grid2-d3.tau-target-T.ci90.1\endcsname{0.219}
\expandafter\def\csname cnt@rev.desc.d1.sspro.p1.Tsealed.grid2-d3.tau-target-T.ci90.1\endcsname{0}
\expandafter\def\csname pct@rev.desc.d1.sspro.p1.Tsealed.grid2-d3.tau-target-T.ci90.1\endcsname{21.9}
\expandafter\def\csname ppt@rev.desc.d1.sspro.p1.Tsealed.grid2-d3.tau-target-T.ci90.1\endcsname{$+$21.9}
\expandafter\def\csname val@rev.desc.d1.sspro.p1.Tsealed.grid2-d3.tau-target-T.ci95-excludes-0\endcsname{1}
\expandafter\def\csname cnt@rev.desc.d1.sspro.p1.Tsealed.grid2-d3.tau-target-T.ci95-excludes-0\endcsname{1}
\expandafter\def\csname pct@rev.desc.d1.sspro.p1.Tsealed.grid2-d3.tau-target-T.ci95-excludes-0\endcsname{100.0}
\expandafter\def\csname ppt@rev.desc.d1.sspro.p1.Tsealed.grid2-d3.tau-target-T.ci95-excludes-0\endcsname{$+$100.0}
\expandafter\def\csname val@rev.desc.d1.sspro.p1.Tsealed.grid2-d3.tau-target-T.ci95.0\endcsname{0.010}
\expandafter\def\csname cnt@rev.desc.d1.sspro.p1.Tsealed.grid2-d3.tau-target-T.ci95.0\endcsname{0}
\expandafter\def\csname pct@rev.desc.d1.sspro.p1.Tsealed.grid2-d3.tau-target-T.ci95.0\endcsname{1.0}
\expandafter\def\csname ppt@rev.desc.d1.sspro.p1.Tsealed.grid2-d3.tau-target-T.ci95.0\endcsname{$+$1.0}
\expandafter\def\csname val@rev.desc.d1.sspro.p1.Tsealed.grid2-d3.tau-target-T.ci95.1\endcsname{0.229}
\expandafter\def\csname cnt@rev.desc.d1.sspro.p1.Tsealed.grid2-d3.tau-target-T.ci95.1\endcsname{0}
\expandafter\def\csname pct@rev.desc.d1.sspro.p1.Tsealed.grid2-d3.tau-target-T.ci95.1\endcsname{22.9}
\expandafter\def\csname ppt@rev.desc.d1.sspro.p1.Tsealed.grid2-d3.tau-target-T.ci95.1\endcsname{$+$22.9}
\expandafter\def\csname val@rev.desc.d1.sspro.p1.Tsealed.grid2-d3.tau-target-T.mean\endcsname{0.115}
\expandafter\def\csname cnt@rev.desc.d1.sspro.p1.Tsealed.grid2-d3.tau-target-T.mean\endcsname{0}
\expandafter\def\csname pct@rev.desc.d1.sspro.p1.Tsealed.grid2-d3.tau-target-T.mean\endcsname{11.5}
\expandafter\def\csname ppt@rev.desc.d1.sspro.p1.Tsealed.grid2-d3.tau-target-T.mean\endcsname{$+$11.5}
\expandafter\def\csname val@rev.desc.d1.sspro.p1.Tsealed.grid2-d3.tau-target-T.n-apps\endcsname{12}
\expandafter\def\csname cnt@rev.desc.d1.sspro.p1.Tsealed.grid2-d3.tau-target-T.n-apps\endcsname{12}
\expandafter\def\csname pct@rev.desc.d1.sspro.p1.Tsealed.grid2-d3.tau-target-T.n-apps\endcsname{1200.0}
\expandafter\def\csname ppt@rev.desc.d1.sspro.p1.Tsealed.grid2-d3.tau-target-T.n-apps\endcsname{$+$1200.0}
\expandafter\def\csname val@rev.desc.d1.sspro.p1.Tsealed.grid2-d3.tau-target-T.tost1.ci90-inside\endcsname{0}
\expandafter\def\csname cnt@rev.desc.d1.sspro.p1.Tsealed.grid2-d3.tau-target-T.tost1.ci90-inside\endcsname{0}
\expandafter\def\csname pct@rev.desc.d1.sspro.p1.Tsealed.grid2-d3.tau-target-T.tost1.ci90-inside\endcsname{0.0}
\expandafter\def\csname ppt@rev.desc.d1.sspro.p1.Tsealed.grid2-d3.tau-target-T.tost1.ci90-inside\endcsname{0.0}
\expandafter\def\csname val@rev.desc.d1.sspro.p1.Tsealed.grid2-d3.tau-target-T.tost1.equivalent\endcsname{0}
\expandafter\def\csname cnt@rev.desc.d1.sspro.p1.Tsealed.grid2-d3.tau-target-T.tost1.equivalent\endcsname{0}
\expandafter\def\csname pct@rev.desc.d1.sspro.p1.Tsealed.grid2-d3.tau-target-T.tost1.equivalent\endcsname{0.0}
\expandafter\def\csname ppt@rev.desc.d1.sspro.p1.Tsealed.grid2-d3.tau-target-T.tost1.equivalent\endcsname{0.0}
\expandafter\def\csname val@rev.desc.d1.sspro.p1.Tsealed.grid2-d3.tau-target-T.tost1.p\endcsname{0.975}
\expandafter\def\csname cnt@rev.desc.d1.sspro.p1.Tsealed.grid2-d3.tau-target-T.tost1.p\endcsname{1}
\expandafter\def\csname pct@rev.desc.d1.sspro.p1.Tsealed.grid2-d3.tau-target-T.tost1.p\endcsname{97.5}
\expandafter\def\csname ppt@rev.desc.d1.sspro.p1.Tsealed.grid2-d3.tau-target-T.tost1.p\endcsname{$+$97.5}
\expandafter\def\csname val@rev.desc.d1.sspro.p1.Tsealed.grid2-d3.tau-target-T.tost5.ci90-inside\endcsname{0}
\expandafter\def\csname cnt@rev.desc.d1.sspro.p1.Tsealed.grid2-d3.tau-target-T.tost5.ci90-inside\endcsname{0}
\expandafter\def\csname pct@rev.desc.d1.sspro.p1.Tsealed.grid2-d3.tau-target-T.tost5.ci90-inside\endcsname{0.0}
\expandafter\def\csname ppt@rev.desc.d1.sspro.p1.Tsealed.grid2-d3.tau-target-T.tost5.ci90-inside\endcsname{0.0}
\expandafter\def\csname val@rev.desc.d1.sspro.p1.Tsealed.grid2-d3.tau-target-T.tost5.equivalent\endcsname{0}
\expandafter\def\csname cnt@rev.desc.d1.sspro.p1.Tsealed.grid2-d3.tau-target-T.tost5.equivalent\endcsname{0}
\expandafter\def\csname pct@rev.desc.d1.sspro.p1.Tsealed.grid2-d3.tau-target-T.tost5.equivalent\endcsname{0.0}
\expandafter\def\csname ppt@rev.desc.d1.sspro.p1.Tsealed.grid2-d3.tau-target-T.tost5.equivalent\endcsname{0.0}
\expandafter\def\csname val@rev.desc.d1.sspro.p1.Tsealed.grid2-d3.tau-target-T.tost5.p\endcsname{0.867}
\expandafter\def\csname cnt@rev.desc.d1.sspro.p1.Tsealed.grid2-d3.tau-target-T.tost5.p\endcsname{1}
\expandafter\def\csname pct@rev.desc.d1.sspro.p1.Tsealed.grid2-d3.tau-target-T.tost5.p\endcsname{86.7}
\expandafter\def\csname ppt@rev.desc.d1.sspro.p1.Tsealed.grid2-d3.tau-target-T.tost5.p\endcsname{$+$86.7}
\expandafter\def\csname val@rev.desc.d1.sspro.p1.Tsealed.grid2-d3.zero-estimate\endcsname{0}
\expandafter\def\csname cnt@rev.desc.d1.sspro.p1.Tsealed.grid2-d3.zero-estimate\endcsname{0}
\expandafter\def\csname pct@rev.desc.d1.sspro.p1.Tsealed.grid2-d3.zero-estimate\endcsname{0.0}
\expandafter\def\csname ppt@rev.desc.d1.sspro.p1.Tsealed.grid2-d3.zero-estimate\endcsname{0.0}
\expandafter\def\csname val@rev.desc.d1.sspro.p1.Tsealed.zoomclick-d3.delta-tau-cross.ci90.0\endcsname{-0.219}
\expandafter\def\csname cnt@rev.desc.d1.sspro.p1.Tsealed.zoomclick-d3.delta-tau-cross.ci90.0\endcsname{0}
\expandafter\def\csname pct@rev.desc.d1.sspro.p1.Tsealed.zoomclick-d3.delta-tau-cross.ci90.0\endcsname{$-$21.9}
\expandafter\def\csname ppt@rev.desc.d1.sspro.p1.Tsealed.zoomclick-d3.delta-tau-cross.ci90.0\endcsname{$-$21.9}
\expandafter\def\csname val@rev.desc.d1.sspro.p1.Tsealed.zoomclick-d3.delta-tau-cross.ci90.1\endcsname{0.073}
\expandafter\def\csname cnt@rev.desc.d1.sspro.p1.Tsealed.zoomclick-d3.delta-tau-cross.ci90.1\endcsname{0}
\expandafter\def\csname pct@rev.desc.d1.sspro.p1.Tsealed.zoomclick-d3.delta-tau-cross.ci90.1\endcsname{7.3}
\expandafter\def\csname ppt@rev.desc.d1.sspro.p1.Tsealed.zoomclick-d3.delta-tau-cross.ci90.1\endcsname{$+$7.3}
\expandafter\def\csname val@rev.desc.d1.sspro.p1.Tsealed.zoomclick-d3.delta-tau-cross.ci95-excludes-0\endcsname{0}
\expandafter\def\csname cnt@rev.desc.d1.sspro.p1.Tsealed.zoomclick-d3.delta-tau-cross.ci95-excludes-0\endcsname{0}
\expandafter\def\csname pct@rev.desc.d1.sspro.p1.Tsealed.zoomclick-d3.delta-tau-cross.ci95-excludes-0\endcsname{0.0}
\expandafter\def\csname ppt@rev.desc.d1.sspro.p1.Tsealed.zoomclick-d3.delta-tau-cross.ci95-excludes-0\endcsname{0.0}
\expandafter\def\csname val@rev.desc.d1.sspro.p1.Tsealed.zoomclick-d3.delta-tau-cross.ci95.0\endcsname{-0.240}
\expandafter\def\csname cnt@rev.desc.d1.sspro.p1.Tsealed.zoomclick-d3.delta-tau-cross.ci95.0\endcsname{0}
\expandafter\def\csname pct@rev.desc.d1.sspro.p1.Tsealed.zoomclick-d3.delta-tau-cross.ci95.0\endcsname{$-$24.0}
\expandafter\def\csname ppt@rev.desc.d1.sspro.p1.Tsealed.zoomclick-d3.delta-tau-cross.ci95.0\endcsname{$-$24.0}
\expandafter\def\csname val@rev.desc.d1.sspro.p1.Tsealed.zoomclick-d3.delta-tau-cross.ci95.1\endcsname{0.094}
\expandafter\def\csname cnt@rev.desc.d1.sspro.p1.Tsealed.zoomclick-d3.delta-tau-cross.ci95.1\endcsname{0}
\expandafter\def\csname pct@rev.desc.d1.sspro.p1.Tsealed.zoomclick-d3.delta-tau-cross.ci95.1\endcsname{9.4}
\expandafter\def\csname ppt@rev.desc.d1.sspro.p1.Tsealed.zoomclick-d3.delta-tau-cross.ci95.1\endcsname{$+$9.4}
\expandafter\def\csname val@rev.desc.d1.sspro.p1.Tsealed.zoomclick-d3.delta-tau-cross.mean\endcsname{-0.073}
\expandafter\def\csname cnt@rev.desc.d1.sspro.p1.Tsealed.zoomclick-d3.delta-tau-cross.mean\endcsname{0}
\expandafter\def\csname pct@rev.desc.d1.sspro.p1.Tsealed.zoomclick-d3.delta-tau-cross.mean\endcsname{$-$7.3}
\expandafter\def\csname ppt@rev.desc.d1.sspro.p1.Tsealed.zoomclick-d3.delta-tau-cross.mean\endcsname{$-$7.3}
\expandafter\def\csname val@rev.desc.d1.sspro.p1.Tsealed.zoomclick-d3.delta-tau-cross.n-apps.0\endcsname{12}
\expandafter\def\csname cnt@rev.desc.d1.sspro.p1.Tsealed.zoomclick-d3.delta-tau-cross.n-apps.0\endcsname{12}
\expandafter\def\csname pct@rev.desc.d1.sspro.p1.Tsealed.zoomclick-d3.delta-tau-cross.n-apps.0\endcsname{1200.0}
\expandafter\def\csname ppt@rev.desc.d1.sspro.p1.Tsealed.zoomclick-d3.delta-tau-cross.n-apps.0\endcsname{$+$1200.0}
\expandafter\def\csname val@rev.desc.d1.sspro.p1.Tsealed.zoomclick-d3.delta-tau-cross.n-apps.1\endcsname{6}
\expandafter\def\csname cnt@rev.desc.d1.sspro.p1.Tsealed.zoomclick-d3.delta-tau-cross.n-apps.1\endcsname{6}
\expandafter\def\csname pct@rev.desc.d1.sspro.p1.Tsealed.zoomclick-d3.delta-tau-cross.n-apps.1\endcsname{600.0}
\expandafter\def\csname ppt@rev.desc.d1.sspro.p1.Tsealed.zoomclick-d3.delta-tau-cross.n-apps.1\endcsname{$+$600.0}
\expandafter\def\csname val@rev.desc.d1.sspro.p1.Tsealed.zoomclick-d3.delta-tau-cross.tost1.ci90-inside\endcsname{0}
\expandafter\def\csname cnt@rev.desc.d1.sspro.p1.Tsealed.zoomclick-d3.delta-tau-cross.tost1.ci90-inside\endcsname{0}
\expandafter\def\csname pct@rev.desc.d1.sspro.p1.Tsealed.zoomclick-d3.delta-tau-cross.tost1.ci90-inside\endcsname{0.0}
\expandafter\def\csname ppt@rev.desc.d1.sspro.p1.Tsealed.zoomclick-d3.delta-tau-cross.tost1.ci90-inside\endcsname{0.0}
\expandafter\def\csname val@rev.desc.d1.sspro.p1.Tsealed.zoomclick-d3.delta-tau-cross.tost1.equivalent\endcsname{0}
\expandafter\def\csname cnt@rev.desc.d1.sspro.p1.Tsealed.zoomclick-d3.delta-tau-cross.tost1.equivalent\endcsname{0}
\expandafter\def\csname pct@rev.desc.d1.sspro.p1.Tsealed.zoomclick-d3.delta-tau-cross.tost1.equivalent\endcsname{0.0}
\expandafter\def\csname ppt@rev.desc.d1.sspro.p1.Tsealed.zoomclick-d3.delta-tau-cross.tost1.equivalent\endcsname{0.0}
\expandafter\def\csname val@rev.desc.d1.sspro.p1.Tsealed.zoomclick-d3.delta-tau-cross.tost1.p\endcsname{0.775}
\expandafter\def\csname cnt@rev.desc.d1.sspro.p1.Tsealed.zoomclick-d3.delta-tau-cross.tost1.p\endcsname{1}
\expandafter\def\csname pct@rev.desc.d1.sspro.p1.Tsealed.zoomclick-d3.delta-tau-cross.tost1.p\endcsname{77.5}
\expandafter\def\csname ppt@rev.desc.d1.sspro.p1.Tsealed.zoomclick-d3.delta-tau-cross.tost1.p\endcsname{$+$77.5}
\expandafter\def\csname val@rev.desc.d1.sspro.p1.Tsealed.zoomclick-d3.delta-tau-cross.tost5.ci90-inside\endcsname{0}
\expandafter\def\csname cnt@rev.desc.d1.sspro.p1.Tsealed.zoomclick-d3.delta-tau-cross.tost5.ci90-inside\endcsname{0}
\expandafter\def\csname pct@rev.desc.d1.sspro.p1.Tsealed.zoomclick-d3.delta-tau-cross.tost5.ci90-inside\endcsname{0.0}
\expandafter\def\csname ppt@rev.desc.d1.sspro.p1.Tsealed.zoomclick-d3.delta-tau-cross.tost5.ci90-inside\endcsname{0.0}
\expandafter\def\csname val@rev.desc.d1.sspro.p1.Tsealed.zoomclick-d3.delta-tau-cross.tost5.equivalent\endcsname{0}
\expandafter\def\csname cnt@rev.desc.d1.sspro.p1.Tsealed.zoomclick-d3.delta-tau-cross.tost5.equivalent\endcsname{0}
\expandafter\def\csname pct@rev.desc.d1.sspro.p1.Tsealed.zoomclick-d3.delta-tau-cross.tost5.equivalent\endcsname{0.0}
\expandafter\def\csname ppt@rev.desc.d1.sspro.p1.Tsealed.zoomclick-d3.delta-tau-cross.tost5.equivalent\endcsname{0.0}
\expandafter\def\csname val@rev.desc.d1.sspro.p1.Tsealed.zoomclick-d3.delta-tau-cross.tost5.p\endcsname{0.609}
\expandafter\def\csname cnt@rev.desc.d1.sspro.p1.Tsealed.zoomclick-d3.delta-tau-cross.tost5.p\endcsname{1}
\expandafter\def\csname pct@rev.desc.d1.sspro.p1.Tsealed.zoomclick-d3.delta-tau-cross.tost5.p\endcsname{60.9}
\expandafter\def\csname ppt@rev.desc.d1.sspro.p1.Tsealed.zoomclick-d3.delta-tau-cross.tost5.p\endcsname{$+$60.9}
\expandafter\def\csname val@rev.desc.d1.sspro.p1.Tsealed.zoomclick-d3.material-flip\endcsname{0}
\expandafter\def\csname cnt@rev.desc.d1.sspro.p1.Tsealed.zoomclick-d3.material-flip\endcsname{0}
\expandafter\def\csname pct@rev.desc.d1.sspro.p1.Tsealed.zoomclick-d3.material-flip\endcsname{0.0}
\expandafter\def\csname ppt@rev.desc.d1.sspro.p1.Tsealed.zoomclick-d3.material-flip\endcsname{0.0}
\expandafter\def\csname val@rev.desc.d1.sspro.p1.Tsealed.zoomclick-d3.n-tasks\endcsname{96}
\expandafter\def\csname cnt@rev.desc.d1.sspro.p1.Tsealed.zoomclick-d3.n-tasks\endcsname{96}
\expandafter\def\csname pct@rev.desc.d1.sspro.p1.Tsealed.zoomclick-d3.n-tasks\endcsname{9600.0}
\expandafter\def\csname ppt@rev.desc.d1.sspro.p1.Tsealed.zoomclick-d3.n-tasks\endcsname{$+$9600.0}
\expandafter\def\csname val@rev.desc.d1.sspro.p1.Tsealed.zoomclick-d3.n-tasks-source-C\endcsname{48}
\expandafter\def\csname cnt@rev.desc.d1.sspro.p1.Tsealed.zoomclick-d3.n-tasks-source-C\endcsname{48}
\expandafter\def\csname pct@rev.desc.d1.sspro.p1.Tsealed.zoomclick-d3.n-tasks-source-C\endcsname{4800.0}
\expandafter\def\csname ppt@rev.desc.d1.sspro.p1.Tsealed.zoomclick-d3.n-tasks-source-C\endcsname{$+$4800.0}
\expandafter\def\csname val@rev.desc.d1.sspro.p1.Tsealed.zoomclick-d3.paper-tau-T\endcsname{0.177}
\expandafter\def\csname cnt@rev.desc.d1.sspro.p1.Tsealed.zoomclick-d3.paper-tau-T\endcsname{0}
\expandafter\def\csname pct@rev.desc.d1.sspro.p1.Tsealed.zoomclick-d3.paper-tau-T\endcsname{17.7}
\expandafter\def\csname ppt@rev.desc.d1.sspro.p1.Tsealed.zoomclick-d3.paper-tau-T\endcsname{$+$17.7}
\expandafter\def\csname val@rev.desc.d1.sspro.p1.Tsealed.zoomclick-d3.reproduces\endcsname{1}
\expandafter\def\csname cnt@rev.desc.d1.sspro.p1.Tsealed.zoomclick-d3.reproduces\endcsname{1}
\expandafter\def\csname pct@rev.desc.d1.sspro.p1.Tsealed.zoomclick-d3.reproduces\endcsname{100.0}
\expandafter\def\csname ppt@rev.desc.d1.sspro.p1.Tsealed.zoomclick-d3.reproduces\endcsname{$+$100.0}
\expandafter\def\csname val@rev.desc.d1.sspro.p1.Tsealed.zoomclick-d3.reproduces-ci\endcsname{1}
\expandafter\def\csname cnt@rev.desc.d1.sspro.p1.Tsealed.zoomclick-d3.reproduces-ci\endcsname{1}
\expandafter\def\csname pct@rev.desc.d1.sspro.p1.Tsealed.zoomclick-d3.reproduces-ci\endcsname{100.0}
\expandafter\def\csname ppt@rev.desc.d1.sspro.p1.Tsealed.zoomclick-d3.reproduces-ci\endcsname{$+$100.0}
\expandafter\def\csname val@rev.desc.d1.sspro.p1.Tsealed.zoomclick-d3.sign-differs\endcsname{0}
\expandafter\def\csname cnt@rev.desc.d1.sspro.p1.Tsealed.zoomclick-d3.sign-differs\endcsname{0}
\expandafter\def\csname pct@rev.desc.d1.sspro.p1.Tsealed.zoomclick-d3.sign-differs\endcsname{0.0}
\expandafter\def\csname ppt@rev.desc.d1.sspro.p1.Tsealed.zoomclick-d3.sign-differs\endcsname{0.0}
\expandafter\def\csname val@rev.desc.d1.sspro.p1.Tsealed.zoomclick-d3.tau-source-C.ci90.0\endcsname{0.125}
\expandafter\def\csname cnt@rev.desc.d1.sspro.p1.Tsealed.zoomclick-d3.tau-source-C.ci90.0\endcsname{0}
\expandafter\def\csname pct@rev.desc.d1.sspro.p1.Tsealed.zoomclick-d3.tau-source-C.ci90.0\endcsname{12.5}
\expandafter\def\csname ppt@rev.desc.d1.sspro.p1.Tsealed.zoomclick-d3.tau-source-C.ci90.0\endcsname{$+$12.5}
\expandafter\def\csname val@rev.desc.d1.sspro.p1.Tsealed.zoomclick-d3.tau-source-C.ci90.1\endcsname{0.375}
\expandafter\def\csname cnt@rev.desc.d1.sspro.p1.Tsealed.zoomclick-d3.tau-source-C.ci90.1\endcsname{0}
\expandafter\def\csname pct@rev.desc.d1.sspro.p1.Tsealed.zoomclick-d3.tau-source-C.ci90.1\endcsname{37.5}
\expandafter\def\csname ppt@rev.desc.d1.sspro.p1.Tsealed.zoomclick-d3.tau-source-C.ci90.1\endcsname{$+$37.5}
\expandafter\def\csname val@rev.desc.d1.sspro.p1.Tsealed.zoomclick-d3.tau-source-C.ci95-excludes-0\endcsname{1}
\expandafter\def\csname cnt@rev.desc.d1.sspro.p1.Tsealed.zoomclick-d3.tau-source-C.ci95-excludes-0\endcsname{1}
\expandafter\def\csname pct@rev.desc.d1.sspro.p1.Tsealed.zoomclick-d3.tau-source-C.ci95-excludes-0\endcsname{100.0}
\expandafter\def\csname ppt@rev.desc.d1.sspro.p1.Tsealed.zoomclick-d3.tau-source-C.ci95-excludes-0\endcsname{$+$100.0}
\expandafter\def\csname val@rev.desc.d1.sspro.p1.Tsealed.zoomclick-d3.tau-source-C.ci95.0\endcsname{0.104}
\expandafter\def\csname cnt@rev.desc.d1.sspro.p1.Tsealed.zoomclick-d3.tau-source-C.ci95.0\endcsname{0}
\expandafter\def\csname pct@rev.desc.d1.sspro.p1.Tsealed.zoomclick-d3.tau-source-C.ci95.0\endcsname{10.4}
\expandafter\def\csname ppt@rev.desc.d1.sspro.p1.Tsealed.zoomclick-d3.tau-source-C.ci95.0\endcsname{$+$10.4}
\expandafter\def\csname val@rev.desc.d1.sspro.p1.Tsealed.zoomclick-d3.tau-source-C.ci95.1\endcsname{0.396}
\expandafter\def\csname cnt@rev.desc.d1.sspro.p1.Tsealed.zoomclick-d3.tau-source-C.ci95.1\endcsname{0}
\expandafter\def\csname pct@rev.desc.d1.sspro.p1.Tsealed.zoomclick-d3.tau-source-C.ci95.1\endcsname{39.6}
\expandafter\def\csname ppt@rev.desc.d1.sspro.p1.Tsealed.zoomclick-d3.tau-source-C.ci95.1\endcsname{$+$39.6}
\expandafter\def\csname val@rev.desc.d1.sspro.p1.Tsealed.zoomclick-d3.tau-source-C.mean\endcsname{0.250}
\expandafter\def\csname cnt@rev.desc.d1.sspro.p1.Tsealed.zoomclick-d3.tau-source-C.mean\endcsname{0}
\expandafter\def\csname pct@rev.desc.d1.sspro.p1.Tsealed.zoomclick-d3.tau-source-C.mean\endcsname{25.0}
\expandafter\def\csname ppt@rev.desc.d1.sspro.p1.Tsealed.zoomclick-d3.tau-source-C.mean\endcsname{$+$25.0}
\expandafter\def\csname val@rev.desc.d1.sspro.p1.Tsealed.zoomclick-d3.tau-source-C.n-apps\endcsname{6}
\expandafter\def\csname cnt@rev.desc.d1.sspro.p1.Tsealed.zoomclick-d3.tau-source-C.n-apps\endcsname{6}
\expandafter\def\csname pct@rev.desc.d1.sspro.p1.Tsealed.zoomclick-d3.tau-source-C.n-apps\endcsname{600.0}
\expandafter\def\csname ppt@rev.desc.d1.sspro.p1.Tsealed.zoomclick-d3.tau-source-C.n-apps\endcsname{$+$600.0}
\expandafter\def\csname val@rev.desc.d1.sspro.p1.Tsealed.zoomclick-d3.tau-source-C.tost1.ci90-inside\endcsname{0}
\expandafter\def\csname cnt@rev.desc.d1.sspro.p1.Tsealed.zoomclick-d3.tau-source-C.tost1.ci90-inside\endcsname{0}
\expandafter\def\csname pct@rev.desc.d1.sspro.p1.Tsealed.zoomclick-d3.tau-source-C.tost1.ci90-inside\endcsname{0.0}
\expandafter\def\csname ppt@rev.desc.d1.sspro.p1.Tsealed.zoomclick-d3.tau-source-C.tost1.ci90-inside\endcsname{0.0}
\expandafter\def\csname val@rev.desc.d1.sspro.p1.Tsealed.zoomclick-d3.tau-source-C.tost1.equivalent\endcsname{0}
\expandafter\def\csname cnt@rev.desc.d1.sspro.p1.Tsealed.zoomclick-d3.tau-source-C.tost1.equivalent\endcsname{0}
\expandafter\def\csname pct@rev.desc.d1.sspro.p1.Tsealed.zoomclick-d3.tau-source-C.tost1.equivalent\endcsname{0.0}
\expandafter\def\csname ppt@rev.desc.d1.sspro.p1.Tsealed.zoomclick-d3.tau-source-C.tost1.equivalent\endcsname{0.0}
\expandafter\def\csname val@rev.desc.d1.sspro.p1.Tsealed.zoomclick-d3.tau-source-C.tost1.p\endcsname{1.000}
\expandafter\def\csname cnt@rev.desc.d1.sspro.p1.Tsealed.zoomclick-d3.tau-source-C.tost1.p\endcsname{1}
\expandafter\def\csname pct@rev.desc.d1.sspro.p1.Tsealed.zoomclick-d3.tau-source-C.tost1.p\endcsname{100.0}
\expandafter\def\csname ppt@rev.desc.d1.sspro.p1.Tsealed.zoomclick-d3.tau-source-C.tost1.p\endcsname{$+$100.0}
\expandafter\def\csname val@rev.desc.d1.sspro.p1.Tsealed.zoomclick-d3.tau-source-C.tost5.ci90-inside\endcsname{0}
\expandafter\def\csname cnt@rev.desc.d1.sspro.p1.Tsealed.zoomclick-d3.tau-source-C.tost5.ci90-inside\endcsname{0}
\expandafter\def\csname pct@rev.desc.d1.sspro.p1.Tsealed.zoomclick-d3.tau-source-C.tost5.ci90-inside\endcsname{0.0}
\expandafter\def\csname ppt@rev.desc.d1.sspro.p1.Tsealed.zoomclick-d3.tau-source-C.tost5.ci90-inside\endcsname{0.0}
\expandafter\def\csname val@rev.desc.d1.sspro.p1.Tsealed.zoomclick-d3.tau-source-C.tost5.equivalent\endcsname{0}
\expandafter\def\csname cnt@rev.desc.d1.sspro.p1.Tsealed.zoomclick-d3.tau-source-C.tost5.equivalent\endcsname{0}
\expandafter\def\csname pct@rev.desc.d1.sspro.p1.Tsealed.zoomclick-d3.tau-source-C.tost5.equivalent\endcsname{0.0}
\expandafter\def\csname ppt@rev.desc.d1.sspro.p1.Tsealed.zoomclick-d3.tau-source-C.tost5.equivalent\endcsname{0.0}
\expandafter\def\csname val@rev.desc.d1.sspro.p1.Tsealed.zoomclick-d3.tau-source-C.tost5.p\endcsname{0.998}
\expandafter\def\csname cnt@rev.desc.d1.sspro.p1.Tsealed.zoomclick-d3.tau-source-C.tost5.p\endcsname{1}
\expandafter\def\csname pct@rev.desc.d1.sspro.p1.Tsealed.zoomclick-d3.tau-source-C.tost5.p\endcsname{99.8}
\expandafter\def\csname ppt@rev.desc.d1.sspro.p1.Tsealed.zoomclick-d3.tau-source-C.tost5.p\endcsname{$+$99.8}
\expandafter\def\csname val@rev.desc.d1.sspro.p1.Tsealed.zoomclick-d3.tau-target-T.ci90.0\endcsname{0.104}
\expandafter\def\csname cnt@rev.desc.d1.sspro.p1.Tsealed.zoomclick-d3.tau-target-T.ci90.0\endcsname{0}
\expandafter\def\csname pct@rev.desc.d1.sspro.p1.Tsealed.zoomclick-d3.tau-target-T.ci90.0\endcsname{10.4}
\expandafter\def\csname ppt@rev.desc.d1.sspro.p1.Tsealed.zoomclick-d3.tau-target-T.ci90.0\endcsname{$+$10.4}
\expandafter\def\csname val@rev.desc.d1.sspro.p1.Tsealed.zoomclick-d3.tau-target-T.ci90.1\endcsname{0.271}
\expandafter\def\csname cnt@rev.desc.d1.sspro.p1.Tsealed.zoomclick-d3.tau-target-T.ci90.1\endcsname{0}
\expandafter\def\csname pct@rev.desc.d1.sspro.p1.Tsealed.zoomclick-d3.tau-target-T.ci90.1\endcsname{27.1}
\expandafter\def\csname ppt@rev.desc.d1.sspro.p1.Tsealed.zoomclick-d3.tau-target-T.ci90.1\endcsname{$+$27.1}
\expandafter\def\csname val@rev.desc.d1.sspro.p1.Tsealed.zoomclick-d3.tau-target-T.ci95-excludes-0\endcsname{1}
\expandafter\def\csname cnt@rev.desc.d1.sspro.p1.Tsealed.zoomclick-d3.tau-target-T.ci95-excludes-0\endcsname{1}
\expandafter\def\csname pct@rev.desc.d1.sspro.p1.Tsealed.zoomclick-d3.tau-target-T.ci95-excludes-0\endcsname{100.0}
\expandafter\def\csname ppt@rev.desc.d1.sspro.p1.Tsealed.zoomclick-d3.tau-target-T.ci95-excludes-0\endcsname{$+$100.0}
\expandafter\def\csname val@rev.desc.d1.sspro.p1.Tsealed.zoomclick-d3.tau-target-T.ci95.0\endcsname{0.094}
\expandafter\def\csname cnt@rev.desc.d1.sspro.p1.Tsealed.zoomclick-d3.tau-target-T.ci95.0\endcsname{0}
\expandafter\def\csname pct@rev.desc.d1.sspro.p1.Tsealed.zoomclick-d3.tau-target-T.ci95.0\endcsname{9.4}
\expandafter\def\csname ppt@rev.desc.d1.sspro.p1.Tsealed.zoomclick-d3.tau-target-T.ci95.0\endcsname{$+$9.4}
\expandafter\def\csname val@rev.desc.d1.sspro.p1.Tsealed.zoomclick-d3.tau-target-T.ci95.1\endcsname{0.292}
\expandafter\def\csname cnt@rev.desc.d1.sspro.p1.Tsealed.zoomclick-d3.tau-target-T.ci95.1\endcsname{0}
\expandafter\def\csname pct@rev.desc.d1.sspro.p1.Tsealed.zoomclick-d3.tau-target-T.ci95.1\endcsname{29.2}
\expandafter\def\csname ppt@rev.desc.d1.sspro.p1.Tsealed.zoomclick-d3.tau-target-T.ci95.1\endcsname{$+$29.2}
\expandafter\def\csname val@rev.desc.d1.sspro.p1.Tsealed.zoomclick-d3.tau-target-T.mean\endcsname{0.177}
\expandafter\def\csname cnt@rev.desc.d1.sspro.p1.Tsealed.zoomclick-d3.tau-target-T.mean\endcsname{0}
\expandafter\def\csname pct@rev.desc.d1.sspro.p1.Tsealed.zoomclick-d3.tau-target-T.mean\endcsname{17.7}
\expandafter\def\csname ppt@rev.desc.d1.sspro.p1.Tsealed.zoomclick-d3.tau-target-T.mean\endcsname{$+$17.7}
\expandafter\def\csname val@rev.desc.d1.sspro.p1.Tsealed.zoomclick-d3.tau-target-T.n-apps\endcsname{12}
\expandafter\def\csname cnt@rev.desc.d1.sspro.p1.Tsealed.zoomclick-d3.tau-target-T.n-apps\endcsname{12}
\expandafter\def\csname pct@rev.desc.d1.sspro.p1.Tsealed.zoomclick-d3.tau-target-T.n-apps\endcsname{1200.0}
\expandafter\def\csname ppt@rev.desc.d1.sspro.p1.Tsealed.zoomclick-d3.tau-target-T.n-apps\endcsname{$+$1200.0}
\expandafter\def\csname val@rev.desc.d1.sspro.p1.Tsealed.zoomclick-d3.tau-target-T.tost1.ci90-inside\endcsname{0}
\expandafter\def\csname cnt@rev.desc.d1.sspro.p1.Tsealed.zoomclick-d3.tau-target-T.tost1.ci90-inside\endcsname{0}
\expandafter\def\csname pct@rev.desc.d1.sspro.p1.Tsealed.zoomclick-d3.tau-target-T.tost1.ci90-inside\endcsname{0.0}
\expandafter\def\csname ppt@rev.desc.d1.sspro.p1.Tsealed.zoomclick-d3.tau-target-T.tost1.ci90-inside\endcsname{0.0}
\expandafter\def\csname val@rev.desc.d1.sspro.p1.Tsealed.zoomclick-d3.tau-target-T.tost1.equivalent\endcsname{0}
\expandafter\def\csname cnt@rev.desc.d1.sspro.p1.Tsealed.zoomclick-d3.tau-target-T.tost1.equivalent\endcsname{0}
\expandafter\def\csname pct@rev.desc.d1.sspro.p1.Tsealed.zoomclick-d3.tau-target-T.tost1.equivalent\endcsname{0.0}
\expandafter\def\csname ppt@rev.desc.d1.sspro.p1.Tsealed.zoomclick-d3.tau-target-T.tost1.equivalent\endcsname{0.0}
\expandafter\def\csname val@rev.desc.d1.sspro.p1.Tsealed.zoomclick-d3.tau-target-T.tost1.p\endcsname{1.000}
\expandafter\def\csname cnt@rev.desc.d1.sspro.p1.Tsealed.zoomclick-d3.tau-target-T.tost1.p\endcsname{1}
\expandafter\def\csname pct@rev.desc.d1.sspro.p1.Tsealed.zoomclick-d3.tau-target-T.tost1.p\endcsname{100.0}
\expandafter\def\csname ppt@rev.desc.d1.sspro.p1.Tsealed.zoomclick-d3.tau-target-T.tost1.p\endcsname{$+$100.0}
\expandafter\def\csname val@rev.desc.d1.sspro.p1.Tsealed.zoomclick-d3.tau-target-T.tost5.ci90-inside\endcsname{0}
\expandafter\def\csname cnt@rev.desc.d1.sspro.p1.Tsealed.zoomclick-d3.tau-target-T.tost5.ci90-inside\endcsname{0}
\expandafter\def\csname pct@rev.desc.d1.sspro.p1.Tsealed.zoomclick-d3.tau-target-T.tost5.ci90-inside\endcsname{0.0}
\expandafter\def\csname ppt@rev.desc.d1.sspro.p1.Tsealed.zoomclick-d3.tau-target-T.tost5.ci90-inside\endcsname{0.0}
\expandafter\def\csname val@rev.desc.d1.sspro.p1.Tsealed.zoomclick-d3.tau-target-T.tost5.equivalent\endcsname{0}
\expandafter\def\csname cnt@rev.desc.d1.sspro.p1.Tsealed.zoomclick-d3.tau-target-T.tost5.equivalent\endcsname{0}
\expandafter\def\csname pct@rev.desc.d1.sspro.p1.Tsealed.zoomclick-d3.tau-target-T.tost5.equivalent\endcsname{0.0}
\expandafter\def\csname ppt@rev.desc.d1.sspro.p1.Tsealed.zoomclick-d3.tau-target-T.tost5.equivalent\endcsname{0.0}
\expandafter\def\csname val@rev.desc.d1.sspro.p1.Tsealed.zoomclick-d3.tau-target-T.tost5.p\endcsname{0.999}
\expandafter\def\csname cnt@rev.desc.d1.sspro.p1.Tsealed.zoomclick-d3.tau-target-T.tost5.p\endcsname{1}
\expandafter\def\csname pct@rev.desc.d1.sspro.p1.Tsealed.zoomclick-d3.tau-target-T.tost5.p\endcsname{99.9}
\expandafter\def\csname ppt@rev.desc.d1.sspro.p1.Tsealed.zoomclick-d3.tau-target-T.tost5.p\endcsname{$+$99.9}
\expandafter\def\csname val@rev.desc.d1.sspro.p1.Tsealed.zoomclick-d3.zero-estimate\endcsname{0}
\expandafter\def\csname cnt@rev.desc.d1.sspro.p1.Tsealed.zoomclick-d3.zero-estimate\endcsname{0}
\expandafter\def\csname pct@rev.desc.d1.sspro.p1.Tsealed.zoomclick-d3.zero-estimate\endcsname{0.0}
\expandafter\def\csname ppt@rev.desc.d1.sspro.p1.Tsealed.zoomclick-d3.zero-estimate\endcsname{0.0}
\expandafter\def\csname val@rev.desc.d1.sspro.p2.C.center-d1.delta-tau.ci90.0\endcsname{-0.146}
\expandafter\def\csname cnt@rev.desc.d1.sspro.p2.C.center-d1.delta-tau.ci90.0\endcsname{0}
\expandafter\def\csname pct@rev.desc.d1.sspro.p2.C.center-d1.delta-tau.ci90.0\endcsname{$-$14.6}
\expandafter\def\csname ppt@rev.desc.d1.sspro.p2.C.center-d1.delta-tau.ci90.0\endcsname{$-$14.6}
\expandafter\def\csname val@rev.desc.d1.sspro.p2.C.center-d1.delta-tau.ci90.1\endcsname{0.021}
\expandafter\def\csname cnt@rev.desc.d1.sspro.p2.C.center-d1.delta-tau.ci90.1\endcsname{0}
\expandafter\def\csname pct@rev.desc.d1.sspro.p2.C.center-d1.delta-tau.ci90.1\endcsname{2.1}
\expandafter\def\csname ppt@rev.desc.d1.sspro.p2.C.center-d1.delta-tau.ci90.1\endcsname{$+$2.1}
\expandafter\def\csname val@rev.desc.d1.sspro.p2.C.center-d1.delta-tau.ci95-excludes-0\endcsname{0}
\expandafter\def\csname cnt@rev.desc.d1.sspro.p2.C.center-d1.delta-tau.ci95-excludes-0\endcsname{0}
\expandafter\def\csname pct@rev.desc.d1.sspro.p2.C.center-d1.delta-tau.ci95-excludes-0\endcsname{0.0}
\expandafter\def\csname ppt@rev.desc.d1.sspro.p2.C.center-d1.delta-tau.ci95-excludes-0\endcsname{0.0}
\expandafter\def\csname val@rev.desc.d1.sspro.p2.C.center-d1.delta-tau.ci95.0\endcsname{-0.146}
\expandafter\def\csname cnt@rev.desc.d1.sspro.p2.C.center-d1.delta-tau.ci95.0\endcsname{0}
\expandafter\def\csname pct@rev.desc.d1.sspro.p2.C.center-d1.delta-tau.ci95.0\endcsname{$-$14.6}
\expandafter\def\csname ppt@rev.desc.d1.sspro.p2.C.center-d1.delta-tau.ci95.0\endcsname{$-$14.6}
\expandafter\def\csname val@rev.desc.d1.sspro.p2.C.center-d1.delta-tau.ci95.1\endcsname{0.042}
\expandafter\def\csname cnt@rev.desc.d1.sspro.p2.C.center-d1.delta-tau.ci95.1\endcsname{0}
\expandafter\def\csname pct@rev.desc.d1.sspro.p2.C.center-d1.delta-tau.ci95.1\endcsname{4.2}
\expandafter\def\csname ppt@rev.desc.d1.sspro.p2.C.center-d1.delta-tau.ci95.1\endcsname{$+$4.2}
\expandafter\def\csname val@rev.desc.d1.sspro.p2.C.center-d1.delta-tau.mean\endcsname{-0.062}
\expandafter\def\csname cnt@rev.desc.d1.sspro.p2.C.center-d1.delta-tau.mean\endcsname{0}
\expandafter\def\csname pct@rev.desc.d1.sspro.p2.C.center-d1.delta-tau.mean\endcsname{$-$6.2}
\expandafter\def\csname ppt@rev.desc.d1.sspro.p2.C.center-d1.delta-tau.mean\endcsname{$-$6.2}
\expandafter\def\csname val@rev.desc.d1.sspro.p2.C.center-d1.delta-tau.n-apps\endcsname{6}
\expandafter\def\csname cnt@rev.desc.d1.sspro.p2.C.center-d1.delta-tau.n-apps\endcsname{6}
\expandafter\def\csname pct@rev.desc.d1.sspro.p2.C.center-d1.delta-tau.n-apps\endcsname{600.0}
\expandafter\def\csname ppt@rev.desc.d1.sspro.p2.C.center-d1.delta-tau.n-apps\endcsname{$+$600.0}
\expandafter\def\csname val@rev.desc.d1.sspro.p2.C.center-d1.delta-tau.tost1.ci90-inside\endcsname{0}
\expandafter\def\csname cnt@rev.desc.d1.sspro.p2.C.center-d1.delta-tau.tost1.ci90-inside\endcsname{0}
\expandafter\def\csname pct@rev.desc.d1.sspro.p2.C.center-d1.delta-tau.tost1.ci90-inside\endcsname{0.0}
\expandafter\def\csname ppt@rev.desc.d1.sspro.p2.C.center-d1.delta-tau.tost1.ci90-inside\endcsname{0.0}
\expandafter\def\csname val@rev.desc.d1.sspro.p2.C.center-d1.delta-tau.tost1.equivalent\endcsname{0}
\expandafter\def\csname cnt@rev.desc.d1.sspro.p2.C.center-d1.delta-tau.tost1.equivalent\endcsname{0}
\expandafter\def\csname pct@rev.desc.d1.sspro.p2.C.center-d1.delta-tau.tost1.equivalent\endcsname{0.0}
\expandafter\def\csname ppt@rev.desc.d1.sspro.p2.C.center-d1.delta-tau.tost1.equivalent\endcsname{0.0}
\expandafter\def\csname val@rev.desc.d1.sspro.p2.C.center-d1.delta-tau.tost1.p\endcsname{0.862}
\expandafter\def\csname cnt@rev.desc.d1.sspro.p2.C.center-d1.delta-tau.tost1.p\endcsname{1}
\expandafter\def\csname pct@rev.desc.d1.sspro.p2.C.center-d1.delta-tau.tost1.p\endcsname{86.2}
\expandafter\def\csname ppt@rev.desc.d1.sspro.p2.C.center-d1.delta-tau.tost1.p\endcsname{$+$86.2}
\expandafter\def\csname val@rev.desc.d1.sspro.p2.C.center-d1.delta-tau.tost5.ci90-inside\endcsname{0}
\expandafter\def\csname cnt@rev.desc.d1.sspro.p2.C.center-d1.delta-tau.tost5.ci90-inside\endcsname{0}
\expandafter\def\csname pct@rev.desc.d1.sspro.p2.C.center-d1.delta-tau.tost5.ci90-inside\endcsname{0.0}
\expandafter\def\csname ppt@rev.desc.d1.sspro.p2.C.center-d1.delta-tau.tost5.ci90-inside\endcsname{0.0}
\expandafter\def\csname val@rev.desc.d1.sspro.p2.C.center-d1.delta-tau.tost5.equivalent\endcsname{0}
\expandafter\def\csname cnt@rev.desc.d1.sspro.p2.C.center-d1.delta-tau.tost5.equivalent\endcsname{0}
\expandafter\def\csname pct@rev.desc.d1.sspro.p2.C.center-d1.delta-tau.tost5.equivalent\endcsname{0.0}
\expandafter\def\csname ppt@rev.desc.d1.sspro.p2.C.center-d1.delta-tau.tost5.equivalent\endcsname{0.0}
\expandafter\def\csname val@rev.desc.d1.sspro.p2.C.center-d1.delta-tau.tost5.p\endcsname{0.580}
\expandafter\def\csname cnt@rev.desc.d1.sspro.p2.C.center-d1.delta-tau.tost5.p\endcsname{1}
\expandafter\def\csname pct@rev.desc.d1.sspro.p2.C.center-d1.delta-tau.tost5.p\endcsname{58.0}
\expandafter\def\csname ppt@rev.desc.d1.sspro.p2.C.center-d1.delta-tau.tost5.p\endcsname{$+$58.0}
\expandafter\def\csname val@rev.desc.d1.sspro.p2.C.center-d1.material-flip\endcsname{0}
\expandafter\def\csname cnt@rev.desc.d1.sspro.p2.C.center-d1.material-flip\endcsname{0}
\expandafter\def\csname pct@rev.desc.d1.sspro.p2.C.center-d1.material-flip\endcsname{0.0}
\expandafter\def\csname ppt@rev.desc.d1.sspro.p2.C.center-d1.material-flip\endcsname{0.0}
\expandafter\def\csname val@rev.desc.d1.sspro.p2.C.center-d1.n-apps\endcsname{6}
\expandafter\def\csname cnt@rev.desc.d1.sspro.p2.C.center-d1.n-apps\endcsname{6}
\expandafter\def\csname pct@rev.desc.d1.sspro.p2.C.center-d1.n-apps\endcsname{600.0}
\expandafter\def\csname ppt@rev.desc.d1.sspro.p2.C.center-d1.n-apps\endcsname{$+$600.0}
\expandafter\def\csname val@rev.desc.d1.sspro.p2.C.center-d1.n-tasks\endcsname{48}
\expandafter\def\csname cnt@rev.desc.d1.sspro.p2.C.center-d1.n-tasks\endcsname{48}
\expandafter\def\csname pct@rev.desc.d1.sspro.p2.C.center-d1.n-tasks\endcsname{4800.0}
\expandafter\def\csname ppt@rev.desc.d1.sspro.p2.C.center-d1.n-tasks\endcsname{$+$4800.0}
\expandafter\def\csname val@rev.desc.d1.sspro.p2.C.center-d1.sign-differs\endcsname{0}
\expandafter\def\csname cnt@rev.desc.d1.sspro.p2.C.center-d1.sign-differs\endcsname{0}
\expandafter\def\csname pct@rev.desc.d1.sspro.p2.C.center-d1.sign-differs\endcsname{0.0}
\expandafter\def\csname ppt@rev.desc.d1.sspro.p2.C.center-d1.sign-differs\endcsname{0.0}
\expandafter\def\csname val@rev.desc.d1.sspro.p2.C.center-d1.tau-source.ci90.0\endcsname{0.062}
\expandafter\def\csname cnt@rev.desc.d1.sspro.p2.C.center-d1.tau-source.ci90.0\endcsname{0}
\expandafter\def\csname pct@rev.desc.d1.sspro.p2.C.center-d1.tau-source.ci90.0\endcsname{6.2}
\expandafter\def\csname ppt@rev.desc.d1.sspro.p2.C.center-d1.tau-source.ci90.0\endcsname{$+$6.2}
\expandafter\def\csname val@rev.desc.d1.sspro.p2.C.center-d1.tau-source.ci90.1\endcsname{0.271}
\expandafter\def\csname cnt@rev.desc.d1.sspro.p2.C.center-d1.tau-source.ci90.1\endcsname{0}
\expandafter\def\csname pct@rev.desc.d1.sspro.p2.C.center-d1.tau-source.ci90.1\endcsname{27.1}
\expandafter\def\csname ppt@rev.desc.d1.sspro.p2.C.center-d1.tau-source.ci90.1\endcsname{$+$27.1}
\expandafter\def\csname val@rev.desc.d1.sspro.p2.C.center-d1.tau-source.ci95-excludes-0\endcsname{1}
\expandafter\def\csname cnt@rev.desc.d1.sspro.p2.C.center-d1.tau-source.ci95-excludes-0\endcsname{1}
\expandafter\def\csname pct@rev.desc.d1.sspro.p2.C.center-d1.tau-source.ci95-excludes-0\endcsname{100.0}
\expandafter\def\csname ppt@rev.desc.d1.sspro.p2.C.center-d1.tau-source.ci95-excludes-0\endcsname{$+$100.0}
\expandafter\def\csname val@rev.desc.d1.sspro.p2.C.center-d1.tau-source.ci95.0\endcsname{0.042}
\expandafter\def\csname cnt@rev.desc.d1.sspro.p2.C.center-d1.tau-source.ci95.0\endcsname{0}
\expandafter\def\csname pct@rev.desc.d1.sspro.p2.C.center-d1.tau-source.ci95.0\endcsname{4.2}
\expandafter\def\csname ppt@rev.desc.d1.sspro.p2.C.center-d1.tau-source.ci95.0\endcsname{$+$4.2}
\expandafter\def\csname val@rev.desc.d1.sspro.p2.C.center-d1.tau-source.ci95.1\endcsname{0.292}
\expandafter\def\csname cnt@rev.desc.d1.sspro.p2.C.center-d1.tau-source.ci95.1\endcsname{0}
\expandafter\def\csname pct@rev.desc.d1.sspro.p2.C.center-d1.tau-source.ci95.1\endcsname{29.2}
\expandafter\def\csname ppt@rev.desc.d1.sspro.p2.C.center-d1.tau-source.ci95.1\endcsname{$+$29.2}
\expandafter\def\csname val@rev.desc.d1.sspro.p2.C.center-d1.tau-source.mean\endcsname{0.167}
\expandafter\def\csname cnt@rev.desc.d1.sspro.p2.C.center-d1.tau-source.mean\endcsname{0}
\expandafter\def\csname pct@rev.desc.d1.sspro.p2.C.center-d1.tau-source.mean\endcsname{16.7}
\expandafter\def\csname ppt@rev.desc.d1.sspro.p2.C.center-d1.tau-source.mean\endcsname{$+$16.7}
\expandafter\def\csname val@rev.desc.d1.sspro.p2.C.center-d1.tau-source.n-apps\endcsname{6}
\expandafter\def\csname cnt@rev.desc.d1.sspro.p2.C.center-d1.tau-source.n-apps\endcsname{6}
\expandafter\def\csname pct@rev.desc.d1.sspro.p2.C.center-d1.tau-source.n-apps\endcsname{600.0}
\expandafter\def\csname ppt@rev.desc.d1.sspro.p2.C.center-d1.tau-source.n-apps\endcsname{$+$600.0}
\expandafter\def\csname val@rev.desc.d1.sspro.p2.C.center-d1.tau-source.tost1.ci90-inside\endcsname{0}
\expandafter\def\csname cnt@rev.desc.d1.sspro.p2.C.center-d1.tau-source.tost1.ci90-inside\endcsname{0}
\expandafter\def\csname pct@rev.desc.d1.sspro.p2.C.center-d1.tau-source.tost1.ci90-inside\endcsname{0.0}
\expandafter\def\csname ppt@rev.desc.d1.sspro.p2.C.center-d1.tau-source.tost1.ci90-inside\endcsname{0.0}
\expandafter\def\csname val@rev.desc.d1.sspro.p2.C.center-d1.tau-source.tost1.equivalent\endcsname{0}
\expandafter\def\csname cnt@rev.desc.d1.sspro.p2.C.center-d1.tau-source.tost1.equivalent\endcsname{0}
\expandafter\def\csname pct@rev.desc.d1.sspro.p2.C.center-d1.tau-source.tost1.equivalent\endcsname{0.0}
\expandafter\def\csname ppt@rev.desc.d1.sspro.p2.C.center-d1.tau-source.tost1.equivalent\endcsname{0.0}
\expandafter\def\csname val@rev.desc.d1.sspro.p2.C.center-d1.tau-source.tost1.p\endcsname{0.999}
\expandafter\def\csname cnt@rev.desc.d1.sspro.p2.C.center-d1.tau-source.tost1.p\endcsname{1}
\expandafter\def\csname pct@rev.desc.d1.sspro.p2.C.center-d1.tau-source.tost1.p\endcsname{99.9}
\expandafter\def\csname ppt@rev.desc.d1.sspro.p2.C.center-d1.tau-source.tost1.p\endcsname{$+$99.9}
\expandafter\def\csname val@rev.desc.d1.sspro.p2.C.center-d1.tau-source.tost5.ci90-inside\endcsname{0}
\expandafter\def\csname cnt@rev.desc.d1.sspro.p2.C.center-d1.tau-source.tost5.ci90-inside\endcsname{0}
\expandafter\def\csname pct@rev.desc.d1.sspro.p2.C.center-d1.tau-source.tost5.ci90-inside\endcsname{0.0}
\expandafter\def\csname ppt@rev.desc.d1.sspro.p2.C.center-d1.tau-source.tost5.ci90-inside\endcsname{0.0}
\expandafter\def\csname val@rev.desc.d1.sspro.p2.C.center-d1.tau-source.tost5.equivalent\endcsname{0}
\expandafter\def\csname cnt@rev.desc.d1.sspro.p2.C.center-d1.tau-source.tost5.equivalent\endcsname{0}
\expandafter\def\csname pct@rev.desc.d1.sspro.p2.C.center-d1.tau-source.tost5.equivalent\endcsname{0.0}
\expandafter\def\csname ppt@rev.desc.d1.sspro.p2.C.center-d1.tau-source.tost5.equivalent\endcsname{0.0}
\expandafter\def\csname val@rev.desc.d1.sspro.p2.C.center-d1.tau-source.tost5.p\endcsname{0.971}
\expandafter\def\csname cnt@rev.desc.d1.sspro.p2.C.center-d1.tau-source.tost5.p\endcsname{1}
\expandafter\def\csname pct@rev.desc.d1.sspro.p2.C.center-d1.tau-source.tost5.p\endcsname{97.1}
\expandafter\def\csname ppt@rev.desc.d1.sspro.p2.C.center-d1.tau-source.tost5.p\endcsname{$+$97.1}
\expandafter\def\csname val@rev.desc.d1.sspro.p2.C.center-d1.tau-target.ci90.0\endcsname{0.042}
\expandafter\def\csname cnt@rev.desc.d1.sspro.p2.C.center-d1.tau-target.ci90.0\endcsname{0}
\expandafter\def\csname pct@rev.desc.d1.sspro.p2.C.center-d1.tau-target.ci90.0\endcsname{4.2}
\expandafter\def\csname ppt@rev.desc.d1.sspro.p2.C.center-d1.tau-target.ci90.0\endcsname{$+$4.2}
\expandafter\def\csname val@rev.desc.d1.sspro.p2.C.center-d1.tau-target.ci90.1\endcsname{0.167}
\expandafter\def\csname cnt@rev.desc.d1.sspro.p2.C.center-d1.tau-target.ci90.1\endcsname{0}
\expandafter\def\csname pct@rev.desc.d1.sspro.p2.C.center-d1.tau-target.ci90.1\endcsname{16.7}
\expandafter\def\csname ppt@rev.desc.d1.sspro.p2.C.center-d1.tau-target.ci90.1\endcsname{$+$16.7}
\expandafter\def\csname val@rev.desc.d1.sspro.p2.C.center-d1.tau-target.ci95-excludes-0\endcsname{1}
\expandafter\def\csname cnt@rev.desc.d1.sspro.p2.C.center-d1.tau-target.ci95-excludes-0\endcsname{1}
\expandafter\def\csname pct@rev.desc.d1.sspro.p2.C.center-d1.tau-target.ci95-excludes-0\endcsname{100.0}
\expandafter\def\csname ppt@rev.desc.d1.sspro.p2.C.center-d1.tau-target.ci95-excludes-0\endcsname{$+$100.0}
\expandafter\def\csname val@rev.desc.d1.sspro.p2.C.center-d1.tau-target.ci95.0\endcsname{0.042}
\expandafter\def\csname cnt@rev.desc.d1.sspro.p2.C.center-d1.tau-target.ci95.0\endcsname{0}
\expandafter\def\csname pct@rev.desc.d1.sspro.p2.C.center-d1.tau-target.ci95.0\endcsname{4.2}
\expandafter\def\csname ppt@rev.desc.d1.sspro.p2.C.center-d1.tau-target.ci95.0\endcsname{$+$4.2}
\expandafter\def\csname val@rev.desc.d1.sspro.p2.C.center-d1.tau-target.ci95.1\endcsname{0.167}
\expandafter\def\csname cnt@rev.desc.d1.sspro.p2.C.center-d1.tau-target.ci95.1\endcsname{0}
\expandafter\def\csname pct@rev.desc.d1.sspro.p2.C.center-d1.tau-target.ci95.1\endcsname{16.7}
\expandafter\def\csname ppt@rev.desc.d1.sspro.p2.C.center-d1.tau-target.ci95.1\endcsname{$+$16.7}
\expandafter\def\csname val@rev.desc.d1.sspro.p2.C.center-d1.tau-target.mean\endcsname{0.104}
\expandafter\def\csname cnt@rev.desc.d1.sspro.p2.C.center-d1.tau-target.mean\endcsname{0}
\expandafter\def\csname pct@rev.desc.d1.sspro.p2.C.center-d1.tau-target.mean\endcsname{10.4}
\expandafter\def\csname ppt@rev.desc.d1.sspro.p2.C.center-d1.tau-target.mean\endcsname{$+$10.4}
\expandafter\def\csname val@rev.desc.d1.sspro.p2.C.center-d1.tau-target.n-apps\endcsname{6}
\expandafter\def\csname cnt@rev.desc.d1.sspro.p2.C.center-d1.tau-target.n-apps\endcsname{6}
\expandafter\def\csname pct@rev.desc.d1.sspro.p2.C.center-d1.tau-target.n-apps\endcsname{600.0}
\expandafter\def\csname ppt@rev.desc.d1.sspro.p2.C.center-d1.tau-target.n-apps\endcsname{$+$600.0}
\expandafter\def\csname val@rev.desc.d1.sspro.p2.C.center-d1.tau-target.tost1.ci90-inside\endcsname{0}
\expandafter\def\csname cnt@rev.desc.d1.sspro.p2.C.center-d1.tau-target.tost1.ci90-inside\endcsname{0}
\expandafter\def\csname pct@rev.desc.d1.sspro.p2.C.center-d1.tau-target.tost1.ci90-inside\endcsname{0.0}
\expandafter\def\csname ppt@rev.desc.d1.sspro.p2.C.center-d1.tau-target.tost1.ci90-inside\endcsname{0.0}
\expandafter\def\csname val@rev.desc.d1.sspro.p2.C.center-d1.tau-target.tost1.equivalent\endcsname{0}
\expandafter\def\csname cnt@rev.desc.d1.sspro.p2.C.center-d1.tau-target.tost1.equivalent\endcsname{0}
\expandafter\def\csname pct@rev.desc.d1.sspro.p2.C.center-d1.tau-target.tost1.equivalent\endcsname{0.0}
\expandafter\def\csname ppt@rev.desc.d1.sspro.p2.C.center-d1.tau-target.tost1.equivalent\endcsname{0.0}
\expandafter\def\csname val@rev.desc.d1.sspro.p2.C.center-d1.tau-target.tost1.p\endcsname{0.998}
\expandafter\def\csname cnt@rev.desc.d1.sspro.p2.C.center-d1.tau-target.tost1.p\endcsname{1}
\expandafter\def\csname pct@rev.desc.d1.sspro.p2.C.center-d1.tau-target.tost1.p\endcsname{99.8}
\expandafter\def\csname ppt@rev.desc.d1.sspro.p2.C.center-d1.tau-target.tost1.p\endcsname{$+$99.8}
\expandafter\def\csname val@rev.desc.d1.sspro.p2.C.center-d1.tau-target.tost5.ci90-inside\endcsname{0}
\expandafter\def\csname cnt@rev.desc.d1.sspro.p2.C.center-d1.tau-target.tost5.ci90-inside\endcsname{0}
\expandafter\def\csname pct@rev.desc.d1.sspro.p2.C.center-d1.tau-target.tost5.ci90-inside\endcsname{0.0}
\expandafter\def\csname ppt@rev.desc.d1.sspro.p2.C.center-d1.tau-target.tost5.ci90-inside\endcsname{0.0}
\expandafter\def\csname val@rev.desc.d1.sspro.p2.C.center-d1.tau-target.tost5.equivalent\endcsname{0}
\expandafter\def\csname cnt@rev.desc.d1.sspro.p2.C.center-d1.tau-target.tost5.equivalent\endcsname{0}
\expandafter\def\csname pct@rev.desc.d1.sspro.p2.C.center-d1.tau-target.tost5.equivalent\endcsname{0.0}
\expandafter\def\csname ppt@rev.desc.d1.sspro.p2.C.center-d1.tau-target.tost5.equivalent\endcsname{0.0}
\expandafter\def\csname val@rev.desc.d1.sspro.p2.C.center-d1.tau-target.tost5.p\endcsname{0.930}
\expandafter\def\csname cnt@rev.desc.d1.sspro.p2.C.center-d1.tau-target.tost5.p\endcsname{1}
\expandafter\def\csname pct@rev.desc.d1.sspro.p2.C.center-d1.tau-target.tost5.p\endcsname{93.0}
\expandafter\def\csname ppt@rev.desc.d1.sspro.p2.C.center-d1.tau-target.tost5.p\endcsname{$+$93.0}
\expandafter\def\csname val@rev.desc.d1.sspro.p2.C.center-d1.zero-estimate\endcsname{0}
\expandafter\def\csname cnt@rev.desc.d1.sspro.p2.C.center-d1.zero-estimate\endcsname{0}
\expandafter\def\csname pct@rev.desc.d1.sspro.p2.C.center-d1.zero-estimate\endcsname{0.0}
\expandafter\def\csname ppt@rev.desc.d1.sspro.p2.C.center-d1.zero-estimate\endcsname{0.0}
\expandafter\def\csname val@rev.desc.d1.sspro.p2.C.context-d1.delta-tau.ci90.0\endcsname{-0.062}
\expandafter\def\csname cnt@rev.desc.d1.sspro.p2.C.context-d1.delta-tau.ci90.0\endcsname{0}
\expandafter\def\csname pct@rev.desc.d1.sspro.p2.C.context-d1.delta-tau.ci90.0\endcsname{$-$6.2}
\expandafter\def\csname ppt@rev.desc.d1.sspro.p2.C.context-d1.delta-tau.ci90.0\endcsname{$-$6.2}
\expandafter\def\csname val@rev.desc.d1.sspro.p2.C.context-d1.delta-tau.ci90.1\endcsname{0.062}
\expandafter\def\csname cnt@rev.desc.d1.sspro.p2.C.context-d1.delta-tau.ci90.1\endcsname{0}
\expandafter\def\csname pct@rev.desc.d1.sspro.p2.C.context-d1.delta-tau.ci90.1\endcsname{6.2}
\expandafter\def\csname ppt@rev.desc.d1.sspro.p2.C.context-d1.delta-tau.ci90.1\endcsname{$+$6.2}
\expandafter\def\csname val@rev.desc.d1.sspro.p2.C.context-d1.delta-tau.ci95-excludes-0\endcsname{0}
\expandafter\def\csname cnt@rev.desc.d1.sspro.p2.C.context-d1.delta-tau.ci95-excludes-0\endcsname{0}
\expandafter\def\csname pct@rev.desc.d1.sspro.p2.C.context-d1.delta-tau.ci95-excludes-0\endcsname{0.0}
\expandafter\def\csname ppt@rev.desc.d1.sspro.p2.C.context-d1.delta-tau.ci95-excludes-0\endcsname{0.0}
\expandafter\def\csname val@rev.desc.d1.sspro.p2.C.context-d1.delta-tau.ci95.0\endcsname{-0.083}
\expandafter\def\csname cnt@rev.desc.d1.sspro.p2.C.context-d1.delta-tau.ci95.0\endcsname{0}
\expandafter\def\csname pct@rev.desc.d1.sspro.p2.C.context-d1.delta-tau.ci95.0\endcsname{$-$8.3}
\expandafter\def\csname ppt@rev.desc.d1.sspro.p2.C.context-d1.delta-tau.ci95.0\endcsname{$-$8.3}
\expandafter\def\csname val@rev.desc.d1.sspro.p2.C.context-d1.delta-tau.ci95.1\endcsname{0.083}
\expandafter\def\csname cnt@rev.desc.d1.sspro.p2.C.context-d1.delta-tau.ci95.1\endcsname{0}
\expandafter\def\csname pct@rev.desc.d1.sspro.p2.C.context-d1.delta-tau.ci95.1\endcsname{8.3}
\expandafter\def\csname ppt@rev.desc.d1.sspro.p2.C.context-d1.delta-tau.ci95.1\endcsname{$+$8.3}
\expandafter\def\csname val@rev.desc.d1.sspro.p2.C.context-d1.delta-tau.mean\endcsname{0.000}
\expandafter\def\csname cnt@rev.desc.d1.sspro.p2.C.context-d1.delta-tau.mean\endcsname{0}
\expandafter\def\csname pct@rev.desc.d1.sspro.p2.C.context-d1.delta-tau.mean\endcsname{0.0}
\expandafter\def\csname ppt@rev.desc.d1.sspro.p2.C.context-d1.delta-tau.mean\endcsname{0.0}
\expandafter\def\csname val@rev.desc.d1.sspro.p2.C.context-d1.delta-tau.n-apps\endcsname{6}
\expandafter\def\csname cnt@rev.desc.d1.sspro.p2.C.context-d1.delta-tau.n-apps\endcsname{6}
\expandafter\def\csname pct@rev.desc.d1.sspro.p2.C.context-d1.delta-tau.n-apps\endcsname{600.0}
\expandafter\def\csname ppt@rev.desc.d1.sspro.p2.C.context-d1.delta-tau.n-apps\endcsname{$+$600.0}
\expandafter\def\csname val@rev.desc.d1.sspro.p2.C.context-d1.delta-tau.tost1.ci90-inside\endcsname{0}
\expandafter\def\csname cnt@rev.desc.d1.sspro.p2.C.context-d1.delta-tau.tost1.ci90-inside\endcsname{0}
\expandafter\def\csname pct@rev.desc.d1.sspro.p2.C.context-d1.delta-tau.tost1.ci90-inside\endcsname{0.0}
\expandafter\def\csname ppt@rev.desc.d1.sspro.p2.C.context-d1.delta-tau.tost1.ci90-inside\endcsname{0.0}
\expandafter\def\csname val@rev.desc.d1.sspro.p2.C.context-d1.delta-tau.tost1.equivalent\endcsname{0}
\expandafter\def\csname cnt@rev.desc.d1.sspro.p2.C.context-d1.delta-tau.tost1.equivalent\endcsname{0}
\expandafter\def\csname pct@rev.desc.d1.sspro.p2.C.context-d1.delta-tau.tost1.equivalent\endcsname{0.0}
\expandafter\def\csname ppt@rev.desc.d1.sspro.p2.C.context-d1.delta-tau.tost1.equivalent\endcsname{0.0}
\expandafter\def\csname val@rev.desc.d1.sspro.p2.C.context-d1.delta-tau.tost1.p\endcsname{0.422}
\expandafter\def\csname cnt@rev.desc.d1.sspro.p2.C.context-d1.delta-tau.tost1.p\endcsname{0}
\expandafter\def\csname pct@rev.desc.d1.sspro.p2.C.context-d1.delta-tau.tost1.p\endcsname{42.2}
\expandafter\def\csname ppt@rev.desc.d1.sspro.p2.C.context-d1.delta-tau.tost1.p\endcsname{$+$42.2}
\expandafter\def\csname val@rev.desc.d1.sspro.p2.C.context-d1.delta-tau.tost5.ci90-inside\endcsname{0}
\expandafter\def\csname cnt@rev.desc.d1.sspro.p2.C.context-d1.delta-tau.tost5.ci90-inside\endcsname{0}
\expandafter\def\csname pct@rev.desc.d1.sspro.p2.C.context-d1.delta-tau.tost5.ci90-inside\endcsname{0.0}
\expandafter\def\csname ppt@rev.desc.d1.sspro.p2.C.context-d1.delta-tau.tost5.ci90-inside\endcsname{0.0}
\expandafter\def\csname val@rev.desc.d1.sspro.p2.C.context-d1.delta-tau.tost5.equivalent\endcsname{0}
\expandafter\def\csname cnt@rev.desc.d1.sspro.p2.C.context-d1.delta-tau.tost5.equivalent\endcsname{0}
\expandafter\def\csname pct@rev.desc.d1.sspro.p2.C.context-d1.delta-tau.tost5.equivalent\endcsname{0.0}
\expandafter\def\csname ppt@rev.desc.d1.sspro.p2.C.context-d1.delta-tau.tost5.equivalent\endcsname{0.0}
\expandafter\def\csname val@rev.desc.d1.sspro.p2.C.context-d1.delta-tau.tost5.p\endcsname{0.113}
\expandafter\def\csname cnt@rev.desc.d1.sspro.p2.C.context-d1.delta-tau.tost5.p\endcsname{0}
\expandafter\def\csname pct@rev.desc.d1.sspro.p2.C.context-d1.delta-tau.tost5.p\endcsname{11.3}
\expandafter\def\csname ppt@rev.desc.d1.sspro.p2.C.context-d1.delta-tau.tost5.p\endcsname{$+$11.3}
\expandafter\def\csname val@rev.desc.d1.sspro.p2.C.context-d1.material-flip\endcsname{0}
\expandafter\def\csname cnt@rev.desc.d1.sspro.p2.C.context-d1.material-flip\endcsname{0}
\expandafter\def\csname pct@rev.desc.d1.sspro.p2.C.context-d1.material-flip\endcsname{0.0}
\expandafter\def\csname ppt@rev.desc.d1.sspro.p2.C.context-d1.material-flip\endcsname{0.0}
\expandafter\def\csname val@rev.desc.d1.sspro.p2.C.context-d1.n-apps\endcsname{6}
\expandafter\def\csname cnt@rev.desc.d1.sspro.p2.C.context-d1.n-apps\endcsname{6}
\expandafter\def\csname pct@rev.desc.d1.sspro.p2.C.context-d1.n-apps\endcsname{600.0}
\expandafter\def\csname ppt@rev.desc.d1.sspro.p2.C.context-d1.n-apps\endcsname{$+$600.0}
\expandafter\def\csname val@rev.desc.d1.sspro.p2.C.context-d1.n-tasks\endcsname{48}
\expandafter\def\csname cnt@rev.desc.d1.sspro.p2.C.context-d1.n-tasks\endcsname{48}
\expandafter\def\csname pct@rev.desc.d1.sspro.p2.C.context-d1.n-tasks\endcsname{4800.0}
\expandafter\def\csname ppt@rev.desc.d1.sspro.p2.C.context-d1.n-tasks\endcsname{$+$4800.0}
\expandafter\def\csname val@rev.desc.d1.sspro.p2.C.context-d1.sign-differs\endcsname{0}
\expandafter\def\csname cnt@rev.desc.d1.sspro.p2.C.context-d1.sign-differs\endcsname{0}
\expandafter\def\csname pct@rev.desc.d1.sspro.p2.C.context-d1.sign-differs\endcsname{0.0}
\expandafter\def\csname ppt@rev.desc.d1.sspro.p2.C.context-d1.sign-differs\endcsname{0.0}
\expandafter\def\csname val@rev.desc.d1.sspro.p2.C.context-d1.tau-source.ci90.0\endcsname{0.042}
\expandafter\def\csname cnt@rev.desc.d1.sspro.p2.C.context-d1.tau-source.ci90.0\endcsname{0}
\expandafter\def\csname pct@rev.desc.d1.sspro.p2.C.context-d1.tau-source.ci90.0\endcsname{4.2}
\expandafter\def\csname ppt@rev.desc.d1.sspro.p2.C.context-d1.tau-source.ci90.0\endcsname{$+$4.2}
\expandafter\def\csname val@rev.desc.d1.sspro.p2.C.context-d1.tau-source.ci90.1\endcsname{0.167}
\expandafter\def\csname cnt@rev.desc.d1.sspro.p2.C.context-d1.tau-source.ci90.1\endcsname{0}
\expandafter\def\csname pct@rev.desc.d1.sspro.p2.C.context-d1.tau-source.ci90.1\endcsname{16.7}
\expandafter\def\csname ppt@rev.desc.d1.sspro.p2.C.context-d1.tau-source.ci90.1\endcsname{$+$16.7}
\expandafter\def\csname val@rev.desc.d1.sspro.p2.C.context-d1.tau-source.ci95-excludes-0\endcsname{1}
\expandafter\def\csname cnt@rev.desc.d1.sspro.p2.C.context-d1.tau-source.ci95-excludes-0\endcsname{1}
\expandafter\def\csname pct@rev.desc.d1.sspro.p2.C.context-d1.tau-source.ci95-excludes-0\endcsname{100.0}
\expandafter\def\csname ppt@rev.desc.d1.sspro.p2.C.context-d1.tau-source.ci95-excludes-0\endcsname{$+$100.0}
\expandafter\def\csname val@rev.desc.d1.sspro.p2.C.context-d1.tau-source.ci95.0\endcsname{0.042}
\expandafter\def\csname cnt@rev.desc.d1.sspro.p2.C.context-d1.tau-source.ci95.0\endcsname{0}
\expandafter\def\csname pct@rev.desc.d1.sspro.p2.C.context-d1.tau-source.ci95.0\endcsname{4.2}
\expandafter\def\csname ppt@rev.desc.d1.sspro.p2.C.context-d1.tau-source.ci95.0\endcsname{$+$4.2}
\expandafter\def\csname val@rev.desc.d1.sspro.p2.C.context-d1.tau-source.ci95.1\endcsname{0.167}
\expandafter\def\csname cnt@rev.desc.d1.sspro.p2.C.context-d1.tau-source.ci95.1\endcsname{0}
\expandafter\def\csname pct@rev.desc.d1.sspro.p2.C.context-d1.tau-source.ci95.1\endcsname{16.7}
\expandafter\def\csname ppt@rev.desc.d1.sspro.p2.C.context-d1.tau-source.ci95.1\endcsname{$+$16.7}
\expandafter\def\csname val@rev.desc.d1.sspro.p2.C.context-d1.tau-source.mean\endcsname{0.104}
\expandafter\def\csname cnt@rev.desc.d1.sspro.p2.C.context-d1.tau-source.mean\endcsname{0}
\expandafter\def\csname pct@rev.desc.d1.sspro.p2.C.context-d1.tau-source.mean\endcsname{10.4}
\expandafter\def\csname ppt@rev.desc.d1.sspro.p2.C.context-d1.tau-source.mean\endcsname{$+$10.4}
\expandafter\def\csname val@rev.desc.d1.sspro.p2.C.context-d1.tau-source.n-apps\endcsname{6}
\expandafter\def\csname cnt@rev.desc.d1.sspro.p2.C.context-d1.tau-source.n-apps\endcsname{6}
\expandafter\def\csname pct@rev.desc.d1.sspro.p2.C.context-d1.tau-source.n-apps\endcsname{600.0}
\expandafter\def\csname ppt@rev.desc.d1.sspro.p2.C.context-d1.tau-source.n-apps\endcsname{$+$600.0}
\expandafter\def\csname val@rev.desc.d1.sspro.p2.C.context-d1.tau-source.tost1.ci90-inside\endcsname{0}
\expandafter\def\csname cnt@rev.desc.d1.sspro.p2.C.context-d1.tau-source.tost1.ci90-inside\endcsname{0}
\expandafter\def\csname pct@rev.desc.d1.sspro.p2.C.context-d1.tau-source.tost1.ci90-inside\endcsname{0.0}
\expandafter\def\csname ppt@rev.desc.d1.sspro.p2.C.context-d1.tau-source.tost1.ci90-inside\endcsname{0.0}
\expandafter\def\csname val@rev.desc.d1.sspro.p2.C.context-d1.tau-source.tost1.equivalent\endcsname{0}
\expandafter\def\csname cnt@rev.desc.d1.sspro.p2.C.context-d1.tau-source.tost1.equivalent\endcsname{0}
\expandafter\def\csname pct@rev.desc.d1.sspro.p2.C.context-d1.tau-source.tost1.equivalent\endcsname{0.0}
\expandafter\def\csname ppt@rev.desc.d1.sspro.p2.C.context-d1.tau-source.tost1.equivalent\endcsname{0.0}
\expandafter\def\csname val@rev.desc.d1.sspro.p2.C.context-d1.tau-source.tost1.p\endcsname{0.999}
\expandafter\def\csname cnt@rev.desc.d1.sspro.p2.C.context-d1.tau-source.tost1.p\endcsname{1}
\expandafter\def\csname pct@rev.desc.d1.sspro.p2.C.context-d1.tau-source.tost1.p\endcsname{99.9}
\expandafter\def\csname ppt@rev.desc.d1.sspro.p2.C.context-d1.tau-source.tost1.p\endcsname{$+$99.9}
\expandafter\def\csname val@rev.desc.d1.sspro.p2.C.context-d1.tau-source.tost5.ci90-inside\endcsname{0}
\expandafter\def\csname cnt@rev.desc.d1.sspro.p2.C.context-d1.tau-source.tost5.ci90-inside\endcsname{0}
\expandafter\def\csname pct@rev.desc.d1.sspro.p2.C.context-d1.tau-source.tost5.ci90-inside\endcsname{0.0}
\expandafter\def\csname ppt@rev.desc.d1.sspro.p2.C.context-d1.tau-source.tost5.ci90-inside\endcsname{0.0}
\expandafter\def\csname val@rev.desc.d1.sspro.p2.C.context-d1.tau-source.tost5.equivalent\endcsname{0}
\expandafter\def\csname cnt@rev.desc.d1.sspro.p2.C.context-d1.tau-source.tost5.equivalent\endcsname{0}
\expandafter\def\csname pct@rev.desc.d1.sspro.p2.C.context-d1.tau-source.tost5.equivalent\endcsname{0.0}
\expandafter\def\csname ppt@rev.desc.d1.sspro.p2.C.context-d1.tau-source.tost5.equivalent\endcsname{0.0}
\expandafter\def\csname val@rev.desc.d1.sspro.p2.C.context-d1.tau-source.tost5.p\endcsname{0.931}
\expandafter\def\csname cnt@rev.desc.d1.sspro.p2.C.context-d1.tau-source.tost5.p\endcsname{1}
\expandafter\def\csname pct@rev.desc.d1.sspro.p2.C.context-d1.tau-source.tost5.p\endcsname{93.1}
\expandafter\def\csname ppt@rev.desc.d1.sspro.p2.C.context-d1.tau-source.tost5.p\endcsname{$+$93.1}
\expandafter\def\csname val@rev.desc.d1.sspro.p2.C.context-d1.tau-target.ci90.0\endcsname{0.042}
\expandafter\def\csname cnt@rev.desc.d1.sspro.p2.C.context-d1.tau-target.ci90.0\endcsname{0}
\expandafter\def\csname pct@rev.desc.d1.sspro.p2.C.context-d1.tau-target.ci90.0\endcsname{4.2}
\expandafter\def\csname ppt@rev.desc.d1.sspro.p2.C.context-d1.tau-target.ci90.0\endcsname{$+$4.2}
\expandafter\def\csname val@rev.desc.d1.sspro.p2.C.context-d1.tau-target.ci90.1\endcsname{0.167}
\expandafter\def\csname cnt@rev.desc.d1.sspro.p2.C.context-d1.tau-target.ci90.1\endcsname{0}
\expandafter\def\csname pct@rev.desc.d1.sspro.p2.C.context-d1.tau-target.ci90.1\endcsname{16.7}
\expandafter\def\csname ppt@rev.desc.d1.sspro.p2.C.context-d1.tau-target.ci90.1\endcsname{$+$16.7}
\expandafter\def\csname val@rev.desc.d1.sspro.p2.C.context-d1.tau-target.ci95-excludes-0\endcsname{1}
\expandafter\def\csname cnt@rev.desc.d1.sspro.p2.C.context-d1.tau-target.ci95-excludes-0\endcsname{1}
\expandafter\def\csname pct@rev.desc.d1.sspro.p2.C.context-d1.tau-target.ci95-excludes-0\endcsname{100.0}
\expandafter\def\csname ppt@rev.desc.d1.sspro.p2.C.context-d1.tau-target.ci95-excludes-0\endcsname{$+$100.0}
\expandafter\def\csname val@rev.desc.d1.sspro.p2.C.context-d1.tau-target.ci95.0\endcsname{0.042}
\expandafter\def\csname cnt@rev.desc.d1.sspro.p2.C.context-d1.tau-target.ci95.0\endcsname{0}
\expandafter\def\csname pct@rev.desc.d1.sspro.p2.C.context-d1.tau-target.ci95.0\endcsname{4.2}
\expandafter\def\csname ppt@rev.desc.d1.sspro.p2.C.context-d1.tau-target.ci95.0\endcsname{$+$4.2}
\expandafter\def\csname val@rev.desc.d1.sspro.p2.C.context-d1.tau-target.ci95.1\endcsname{0.167}
\expandafter\def\csname cnt@rev.desc.d1.sspro.p2.C.context-d1.tau-target.ci95.1\endcsname{0}
\expandafter\def\csname pct@rev.desc.d1.sspro.p2.C.context-d1.tau-target.ci95.1\endcsname{16.7}
\expandafter\def\csname ppt@rev.desc.d1.sspro.p2.C.context-d1.tau-target.ci95.1\endcsname{$+$16.7}
\expandafter\def\csname val@rev.desc.d1.sspro.p2.C.context-d1.tau-target.mean\endcsname{0.104}
\expandafter\def\csname cnt@rev.desc.d1.sspro.p2.C.context-d1.tau-target.mean\endcsname{0}
\expandafter\def\csname pct@rev.desc.d1.sspro.p2.C.context-d1.tau-target.mean\endcsname{10.4}
\expandafter\def\csname ppt@rev.desc.d1.sspro.p2.C.context-d1.tau-target.mean\endcsname{$+$10.4}
\expandafter\def\csname val@rev.desc.d1.sspro.p2.C.context-d1.tau-target.n-apps\endcsname{6}
\expandafter\def\csname cnt@rev.desc.d1.sspro.p2.C.context-d1.tau-target.n-apps\endcsname{6}
\expandafter\def\csname pct@rev.desc.d1.sspro.p2.C.context-d1.tau-target.n-apps\endcsname{600.0}
\expandafter\def\csname ppt@rev.desc.d1.sspro.p2.C.context-d1.tau-target.n-apps\endcsname{$+$600.0}
\expandafter\def\csname val@rev.desc.d1.sspro.p2.C.context-d1.tau-target.tost1.ci90-inside\endcsname{0}
\expandafter\def\csname cnt@rev.desc.d1.sspro.p2.C.context-d1.tau-target.tost1.ci90-inside\endcsname{0}
\expandafter\def\csname pct@rev.desc.d1.sspro.p2.C.context-d1.tau-target.tost1.ci90-inside\endcsname{0.0}
\expandafter\def\csname ppt@rev.desc.d1.sspro.p2.C.context-d1.tau-target.tost1.ci90-inside\endcsname{0.0}
\expandafter\def\csname val@rev.desc.d1.sspro.p2.C.context-d1.tau-target.tost1.equivalent\endcsname{0}
\expandafter\def\csname cnt@rev.desc.d1.sspro.p2.C.context-d1.tau-target.tost1.equivalent\endcsname{0}
\expandafter\def\csname pct@rev.desc.d1.sspro.p2.C.context-d1.tau-target.tost1.equivalent\endcsname{0.0}
\expandafter\def\csname ppt@rev.desc.d1.sspro.p2.C.context-d1.tau-target.tost1.equivalent\endcsname{0.0}
\expandafter\def\csname val@rev.desc.d1.sspro.p2.C.context-d1.tau-target.tost1.p\endcsname{0.998}
\expandafter\def\csname cnt@rev.desc.d1.sspro.p2.C.context-d1.tau-target.tost1.p\endcsname{1}
\expandafter\def\csname pct@rev.desc.d1.sspro.p2.C.context-d1.tau-target.tost1.p\endcsname{99.8}
\expandafter\def\csname ppt@rev.desc.d1.sspro.p2.C.context-d1.tau-target.tost1.p\endcsname{$+$99.8}
\expandafter\def\csname val@rev.desc.d1.sspro.p2.C.context-d1.tau-target.tost5.ci90-inside\endcsname{0}
\expandafter\def\csname cnt@rev.desc.d1.sspro.p2.C.context-d1.tau-target.tost5.ci90-inside\endcsname{0}
\expandafter\def\csname pct@rev.desc.d1.sspro.p2.C.context-d1.tau-target.tost5.ci90-inside\endcsname{0.0}
\expandafter\def\csname ppt@rev.desc.d1.sspro.p2.C.context-d1.tau-target.tost5.ci90-inside\endcsname{0.0}
\expandafter\def\csname val@rev.desc.d1.sspro.p2.C.context-d1.tau-target.tost5.equivalent\endcsname{0}
\expandafter\def\csname cnt@rev.desc.d1.sspro.p2.C.context-d1.tau-target.tost5.equivalent\endcsname{0}
\expandafter\def\csname pct@rev.desc.d1.sspro.p2.C.context-d1.tau-target.tost5.equivalent\endcsname{0.0}
\expandafter\def\csname ppt@rev.desc.d1.sspro.p2.C.context-d1.tau-target.tost5.equivalent\endcsname{0.0}
\expandafter\def\csname val@rev.desc.d1.sspro.p2.C.context-d1.tau-target.tost5.p\endcsname{0.930}
\expandafter\def\csname cnt@rev.desc.d1.sspro.p2.C.context-d1.tau-target.tost5.p\endcsname{1}
\expandafter\def\csname pct@rev.desc.d1.sspro.p2.C.context-d1.tau-target.tost5.p\endcsname{93.0}
\expandafter\def\csname ppt@rev.desc.d1.sspro.p2.C.context-d1.tau-target.tost5.p\endcsname{$+$93.0}
\expandafter\def\csname val@rev.desc.d1.sspro.p2.C.context-d1.zero-estimate\endcsname{0}
\expandafter\def\csname cnt@rev.desc.d1.sspro.p2.C.context-d1.zero-estimate\endcsname{0}
\expandafter\def\csname pct@rev.desc.d1.sspro.p2.C.context-d1.zero-estimate\endcsname{0.0}
\expandafter\def\csname ppt@rev.desc.d1.sspro.p2.C.context-d1.zero-estimate\endcsname{0.0}
\expandafter\def\csname val@rev.desc.d1.sspro.p2.C.dual-d2.delta-tau.ci90.0\endcsname{-0.188}
\expandafter\def\csname cnt@rev.desc.d1.sspro.p2.C.dual-d2.delta-tau.ci90.0\endcsname{0}
\expandafter\def\csname pct@rev.desc.d1.sspro.p2.C.dual-d2.delta-tau.ci90.0\endcsname{$-$18.8}
\expandafter\def\csname ppt@rev.desc.d1.sspro.p2.C.dual-d2.delta-tau.ci90.0\endcsname{$-$18.8}
\expandafter\def\csname val@rev.desc.d1.sspro.p2.C.dual-d2.delta-tau.ci90.1\endcsname{0.042}
\expandafter\def\csname cnt@rev.desc.d1.sspro.p2.C.dual-d2.delta-tau.ci90.1\endcsname{0}
\expandafter\def\csname pct@rev.desc.d1.sspro.p2.C.dual-d2.delta-tau.ci90.1\endcsname{4.2}
\expandafter\def\csname ppt@rev.desc.d1.sspro.p2.C.dual-d2.delta-tau.ci90.1\endcsname{$+$4.2}
\expandafter\def\csname val@rev.desc.d1.sspro.p2.C.dual-d2.delta-tau.ci95-excludes-0\endcsname{0}
\expandafter\def\csname cnt@rev.desc.d1.sspro.p2.C.dual-d2.delta-tau.ci95-excludes-0\endcsname{0}
\expandafter\def\csname pct@rev.desc.d1.sspro.p2.C.dual-d2.delta-tau.ci95-excludes-0\endcsname{0.0}
\expandafter\def\csname ppt@rev.desc.d1.sspro.p2.C.dual-d2.delta-tau.ci95-excludes-0\endcsname{0.0}
\expandafter\def\csname val@rev.desc.d1.sspro.p2.C.dual-d2.delta-tau.ci95.0\endcsname{-0.208}
\expandafter\def\csname cnt@rev.desc.d1.sspro.p2.C.dual-d2.delta-tau.ci95.0\endcsname{0}
\expandafter\def\csname pct@rev.desc.d1.sspro.p2.C.dual-d2.delta-tau.ci95.0\endcsname{$-$20.8}
\expandafter\def\csname ppt@rev.desc.d1.sspro.p2.C.dual-d2.delta-tau.ci95.0\endcsname{$-$20.8}
\expandafter\def\csname val@rev.desc.d1.sspro.p2.C.dual-d2.delta-tau.ci95.1\endcsname{0.062}
\expandafter\def\csname cnt@rev.desc.d1.sspro.p2.C.dual-d2.delta-tau.ci95.1\endcsname{0}
\expandafter\def\csname pct@rev.desc.d1.sspro.p2.C.dual-d2.delta-tau.ci95.1\endcsname{6.2}
\expandafter\def\csname ppt@rev.desc.d1.sspro.p2.C.dual-d2.delta-tau.ci95.1\endcsname{$+$6.2}
\expandafter\def\csname val@rev.desc.d1.sspro.p2.C.dual-d2.delta-tau.mean\endcsname{-0.062}
\expandafter\def\csname cnt@rev.desc.d1.sspro.p2.C.dual-d2.delta-tau.mean\endcsname{0}
\expandafter\def\csname pct@rev.desc.d1.sspro.p2.C.dual-d2.delta-tau.mean\endcsname{$-$6.2}
\expandafter\def\csname ppt@rev.desc.d1.sspro.p2.C.dual-d2.delta-tau.mean\endcsname{$-$6.2}
\expandafter\def\csname val@rev.desc.d1.sspro.p2.C.dual-d2.delta-tau.n-apps\endcsname{6}
\expandafter\def\csname cnt@rev.desc.d1.sspro.p2.C.dual-d2.delta-tau.n-apps\endcsname{6}
\expandafter\def\csname pct@rev.desc.d1.sspro.p2.C.dual-d2.delta-tau.n-apps\endcsname{600.0}
\expandafter\def\csname ppt@rev.desc.d1.sspro.p2.C.dual-d2.delta-tau.n-apps\endcsname{$+$600.0}
\expandafter\def\csname val@rev.desc.d1.sspro.p2.C.dual-d2.delta-tau.tost1.ci90-inside\endcsname{0}
\expandafter\def\csname cnt@rev.desc.d1.sspro.p2.C.dual-d2.delta-tau.tost1.ci90-inside\endcsname{0}
\expandafter\def\csname pct@rev.desc.d1.sspro.p2.C.dual-d2.delta-tau.tost1.ci90-inside\endcsname{0.0}
\expandafter\def\csname ppt@rev.desc.d1.sspro.p2.C.dual-d2.delta-tau.tost1.ci90-inside\endcsname{0.0}
\expandafter\def\csname val@rev.desc.d1.sspro.p2.C.dual-d2.delta-tau.tost1.equivalent\endcsname{0}
\expandafter\def\csname cnt@rev.desc.d1.sspro.p2.C.dual-d2.delta-tau.tost1.equivalent\endcsname{0}
\expandafter\def\csname pct@rev.desc.d1.sspro.p2.C.dual-d2.delta-tau.tost1.equivalent\endcsname{0.0}
\expandafter\def\csname ppt@rev.desc.d1.sspro.p2.C.dual-d2.delta-tau.tost1.equivalent\endcsname{0.0}
\expandafter\def\csname val@rev.desc.d1.sspro.p2.C.dual-d2.delta-tau.tost1.p\endcsname{0.754}
\expandafter\def\csname cnt@rev.desc.d1.sspro.p2.C.dual-d2.delta-tau.tost1.p\endcsname{1}
\expandafter\def\csname pct@rev.desc.d1.sspro.p2.C.dual-d2.delta-tau.tost1.p\endcsname{75.4}
\expandafter\def\csname ppt@rev.desc.d1.sspro.p2.C.dual-d2.delta-tau.tost1.p\endcsname{$+$75.4}
\expandafter\def\csname val@rev.desc.d1.sspro.p2.C.dual-d2.delta-tau.tost5.ci90-inside\endcsname{0}
\expandafter\def\csname cnt@rev.desc.d1.sspro.p2.C.dual-d2.delta-tau.tost5.ci90-inside\endcsname{0}
\expandafter\def\csname pct@rev.desc.d1.sspro.p2.C.dual-d2.delta-tau.tost5.ci90-inside\endcsname{0.0}
\expandafter\def\csname ppt@rev.desc.d1.sspro.p2.C.dual-d2.delta-tau.tost5.ci90-inside\endcsname{0.0}
\expandafter\def\csname val@rev.desc.d1.sspro.p2.C.dual-d2.delta-tau.tost5.equivalent\endcsname{0}
\expandafter\def\csname cnt@rev.desc.d1.sspro.p2.C.dual-d2.delta-tau.tost5.equivalent\endcsname{0}
\expandafter\def\csname pct@rev.desc.d1.sspro.p2.C.dual-d2.delta-tau.tost5.equivalent\endcsname{0.0}
\expandafter\def\csname ppt@rev.desc.d1.sspro.p2.C.dual-d2.delta-tau.tost5.equivalent\endcsname{0.0}
\expandafter\def\csname val@rev.desc.d1.sspro.p2.C.dual-d2.delta-tau.tost5.p\endcsname{0.539}
\expandafter\def\csname cnt@rev.desc.d1.sspro.p2.C.dual-d2.delta-tau.tost5.p\endcsname{1}
\expandafter\def\csname pct@rev.desc.d1.sspro.p2.C.dual-d2.delta-tau.tost5.p\endcsname{53.9}
\expandafter\def\csname ppt@rev.desc.d1.sspro.p2.C.dual-d2.delta-tau.tost5.p\endcsname{$+$53.9}
\expandafter\def\csname val@rev.desc.d1.sspro.p2.C.dual-d2.material-flip\endcsname{0}
\expandafter\def\csname cnt@rev.desc.d1.sspro.p2.C.dual-d2.material-flip\endcsname{0}
\expandafter\def\csname pct@rev.desc.d1.sspro.p2.C.dual-d2.material-flip\endcsname{0.0}
\expandafter\def\csname ppt@rev.desc.d1.sspro.p2.C.dual-d2.material-flip\endcsname{0.0}
\expandafter\def\csname val@rev.desc.d1.sspro.p2.C.dual-d2.n-apps\endcsname{6}
\expandafter\def\csname cnt@rev.desc.d1.sspro.p2.C.dual-d2.n-apps\endcsname{6}
\expandafter\def\csname pct@rev.desc.d1.sspro.p2.C.dual-d2.n-apps\endcsname{600.0}
\expandafter\def\csname ppt@rev.desc.d1.sspro.p2.C.dual-d2.n-apps\endcsname{$+$600.0}
\expandafter\def\csname val@rev.desc.d1.sspro.p2.C.dual-d2.n-tasks\endcsname{48}
\expandafter\def\csname cnt@rev.desc.d1.sspro.p2.C.dual-d2.n-tasks\endcsname{48}
\expandafter\def\csname pct@rev.desc.d1.sspro.p2.C.dual-d2.n-tasks\endcsname{4800.0}
\expandafter\def\csname ppt@rev.desc.d1.sspro.p2.C.dual-d2.n-tasks\endcsname{$+$4800.0}
\expandafter\def\csname val@rev.desc.d1.sspro.p2.C.dual-d2.sign-differs\endcsname{0}
\expandafter\def\csname cnt@rev.desc.d1.sspro.p2.C.dual-d2.sign-differs\endcsname{0}
\expandafter\def\csname pct@rev.desc.d1.sspro.p2.C.dual-d2.sign-differs\endcsname{0.0}
\expandafter\def\csname ppt@rev.desc.d1.sspro.p2.C.dual-d2.sign-differs\endcsname{0.0}
\expandafter\def\csname val@rev.desc.d1.sspro.p2.C.dual-d2.tau-source.ci90.0\endcsname{0.021}
\expandafter\def\csname cnt@rev.desc.d1.sspro.p2.C.dual-d2.tau-source.ci90.0\endcsname{0}
\expandafter\def\csname pct@rev.desc.d1.sspro.p2.C.dual-d2.tau-source.ci90.0\endcsname{2.1}
\expandafter\def\csname ppt@rev.desc.d1.sspro.p2.C.dual-d2.tau-source.ci90.0\endcsname{$+$2.1}
\expandafter\def\csname val@rev.desc.d1.sspro.p2.C.dual-d2.tau-source.ci90.1\endcsname{0.229}
\expandafter\def\csname cnt@rev.desc.d1.sspro.p2.C.dual-d2.tau-source.ci90.1\endcsname{0}
\expandafter\def\csname pct@rev.desc.d1.sspro.p2.C.dual-d2.tau-source.ci90.1\endcsname{22.9}
\expandafter\def\csname ppt@rev.desc.d1.sspro.p2.C.dual-d2.tau-source.ci90.1\endcsname{$+$22.9}
\expandafter\def\csname val@rev.desc.d1.sspro.p2.C.dual-d2.tau-source.ci95-excludes-0\endcsname{0}
\expandafter\def\csname cnt@rev.desc.d1.sspro.p2.C.dual-d2.tau-source.ci95-excludes-0\endcsname{0}
\expandafter\def\csname pct@rev.desc.d1.sspro.p2.C.dual-d2.tau-source.ci95-excludes-0\endcsname{0.0}
\expandafter\def\csname ppt@rev.desc.d1.sspro.p2.C.dual-d2.tau-source.ci95-excludes-0\endcsname{0.0}
\expandafter\def\csname val@rev.desc.d1.sspro.p2.C.dual-d2.tau-source.ci95.0\endcsname{0.000}
\expandafter\def\csname cnt@rev.desc.d1.sspro.p2.C.dual-d2.tau-source.ci95.0\endcsname{0}
\expandafter\def\csname pct@rev.desc.d1.sspro.p2.C.dual-d2.tau-source.ci95.0\endcsname{0.0}
\expandafter\def\csname ppt@rev.desc.d1.sspro.p2.C.dual-d2.tau-source.ci95.0\endcsname{0.0}
\expandafter\def\csname val@rev.desc.d1.sspro.p2.C.dual-d2.tau-source.ci95.1\endcsname{0.250}
\expandafter\def\csname cnt@rev.desc.d1.sspro.p2.C.dual-d2.tau-source.ci95.1\endcsname{0}
\expandafter\def\csname pct@rev.desc.d1.sspro.p2.C.dual-d2.tau-source.ci95.1\endcsname{25.0}
\expandafter\def\csname ppt@rev.desc.d1.sspro.p2.C.dual-d2.tau-source.ci95.1\endcsname{$+$25.0}
\expandafter\def\csname val@rev.desc.d1.sspro.p2.C.dual-d2.tau-source.mean\endcsname{0.125}
\expandafter\def\csname cnt@rev.desc.d1.sspro.p2.C.dual-d2.tau-source.mean\endcsname{0}
\expandafter\def\csname pct@rev.desc.d1.sspro.p2.C.dual-d2.tau-source.mean\endcsname{12.5}
\expandafter\def\csname ppt@rev.desc.d1.sspro.p2.C.dual-d2.tau-source.mean\endcsname{$+$12.5}
\expandafter\def\csname val@rev.desc.d1.sspro.p2.C.dual-d2.tau-source.n-apps\endcsname{6}
\expandafter\def\csname cnt@rev.desc.d1.sspro.p2.C.dual-d2.tau-source.n-apps\endcsname{6}
\expandafter\def\csname pct@rev.desc.d1.sspro.p2.C.dual-d2.tau-source.n-apps\endcsname{600.0}
\expandafter\def\csname ppt@rev.desc.d1.sspro.p2.C.dual-d2.tau-source.n-apps\endcsname{$+$600.0}
\expandafter\def\csname val@rev.desc.d1.sspro.p2.C.dual-d2.tau-source.tost1.ci90-inside\endcsname{0}
\expandafter\def\csname cnt@rev.desc.d1.sspro.p2.C.dual-d2.tau-source.tost1.ci90-inside\endcsname{0}
\expandafter\def\csname pct@rev.desc.d1.sspro.p2.C.dual-d2.tau-source.tost1.ci90-inside\endcsname{0.0}
\expandafter\def\csname ppt@rev.desc.d1.sspro.p2.C.dual-d2.tau-source.tost1.ci90-inside\endcsname{0.0}
\expandafter\def\csname val@rev.desc.d1.sspro.p2.C.dual-d2.tau-source.tost1.equivalent\endcsname{0}
\expandafter\def\csname cnt@rev.desc.d1.sspro.p2.C.dual-d2.tau-source.tost1.equivalent\endcsname{0}
\expandafter\def\csname pct@rev.desc.d1.sspro.p2.C.dual-d2.tau-source.tost1.equivalent\endcsname{0.0}
\expandafter\def\csname ppt@rev.desc.d1.sspro.p2.C.dual-d2.tau-source.tost1.equivalent\endcsname{0.0}
\expandafter\def\csname val@rev.desc.d1.sspro.p2.C.dual-d2.tau-source.tost1.p\endcsname{0.958}
\expandafter\def\csname cnt@rev.desc.d1.sspro.p2.C.dual-d2.tau-source.tost1.p\endcsname{1}
\expandafter\def\csname pct@rev.desc.d1.sspro.p2.C.dual-d2.tau-source.tost1.p\endcsname{95.8}
\expandafter\def\csname ppt@rev.desc.d1.sspro.p2.C.dual-d2.tau-source.tost1.p\endcsname{$+$95.8}
\expandafter\def\csname val@rev.desc.d1.sspro.p2.C.dual-d2.tau-source.tost5.ci90-inside\endcsname{0}
\expandafter\def\csname cnt@rev.desc.d1.sspro.p2.C.dual-d2.tau-source.tost5.ci90-inside\endcsname{0}
\expandafter\def\csname pct@rev.desc.d1.sspro.p2.C.dual-d2.tau-source.tost5.ci90-inside\endcsname{0.0}
\expandafter\def\csname ppt@rev.desc.d1.sspro.p2.C.dual-d2.tau-source.tost5.ci90-inside\endcsname{0.0}
\expandafter\def\csname val@rev.desc.d1.sspro.p2.C.dual-d2.tau-source.tost5.equivalent\endcsname{0}
\expandafter\def\csname cnt@rev.desc.d1.sspro.p2.C.dual-d2.tau-source.tost5.equivalent\endcsname{0}
\expandafter\def\csname pct@rev.desc.d1.sspro.p2.C.dual-d2.tau-source.tost5.equivalent\endcsname{0.0}
\expandafter\def\csname ppt@rev.desc.d1.sspro.p2.C.dual-d2.tau-source.tost5.equivalent\endcsname{0.0}
\expandafter\def\csname val@rev.desc.d1.sspro.p2.C.dual-d2.tau-source.tost5.p\endcsname{0.857}
\expandafter\def\csname cnt@rev.desc.d1.sspro.p2.C.dual-d2.tau-source.tost5.p\endcsname{1}
\expandafter\def\csname pct@rev.desc.d1.sspro.p2.C.dual-d2.tau-source.tost5.p\endcsname{85.7}
\expandafter\def\csname ppt@rev.desc.d1.sspro.p2.C.dual-d2.tau-source.tost5.p\endcsname{$+$85.7}
\expandafter\def\csname val@rev.desc.d1.sspro.p2.C.dual-d2.tau-target.ci90.0\endcsname{-0.021}
\expandafter\def\csname cnt@rev.desc.d1.sspro.p2.C.dual-d2.tau-target.ci90.0\endcsname{0}
\expandafter\def\csname pct@rev.desc.d1.sspro.p2.C.dual-d2.tau-target.ci90.0\endcsname{$-$2.1}
\expandafter\def\csname ppt@rev.desc.d1.sspro.p2.C.dual-d2.tau-target.ci90.0\endcsname{$-$2.1}
\expandafter\def\csname val@rev.desc.d1.sspro.p2.C.dual-d2.tau-target.ci90.1\endcsname{0.146}
\expandafter\def\csname cnt@rev.desc.d1.sspro.p2.C.dual-d2.tau-target.ci90.1\endcsname{0}
\expandafter\def\csname pct@rev.desc.d1.sspro.p2.C.dual-d2.tau-target.ci90.1\endcsname{14.6}
\expandafter\def\csname ppt@rev.desc.d1.sspro.p2.C.dual-d2.tau-target.ci90.1\endcsname{$+$14.6}
\expandafter\def\csname val@rev.desc.d1.sspro.p2.C.dual-d2.tau-target.ci95-excludes-0\endcsname{0}
\expandafter\def\csname cnt@rev.desc.d1.sspro.p2.C.dual-d2.tau-target.ci95-excludes-0\endcsname{0}
\expandafter\def\csname pct@rev.desc.d1.sspro.p2.C.dual-d2.tau-target.ci95-excludes-0\endcsname{0.0}
\expandafter\def\csname ppt@rev.desc.d1.sspro.p2.C.dual-d2.tau-target.ci95-excludes-0\endcsname{0.0}
\expandafter\def\csname val@rev.desc.d1.sspro.p2.C.dual-d2.tau-target.ci95.0\endcsname{-0.042}
\expandafter\def\csname cnt@rev.desc.d1.sspro.p2.C.dual-d2.tau-target.ci95.0\endcsname{0}
\expandafter\def\csname pct@rev.desc.d1.sspro.p2.C.dual-d2.tau-target.ci95.0\endcsname{$-$4.2}
\expandafter\def\csname ppt@rev.desc.d1.sspro.p2.C.dual-d2.tau-target.ci95.0\endcsname{$-$4.2}
\expandafter\def\csname val@rev.desc.d1.sspro.p2.C.dual-d2.tau-target.ci95.1\endcsname{0.167}
\expandafter\def\csname cnt@rev.desc.d1.sspro.p2.C.dual-d2.tau-target.ci95.1\endcsname{0}
\expandafter\def\csname pct@rev.desc.d1.sspro.p2.C.dual-d2.tau-target.ci95.1\endcsname{16.7}
\expandafter\def\csname ppt@rev.desc.d1.sspro.p2.C.dual-d2.tau-target.ci95.1\endcsname{$+$16.7}
\expandafter\def\csname val@rev.desc.d1.sspro.p2.C.dual-d2.tau-target.mean\endcsname{0.062}
\expandafter\def\csname cnt@rev.desc.d1.sspro.p2.C.dual-d2.tau-target.mean\endcsname{0}
\expandafter\def\csname pct@rev.desc.d1.sspro.p2.C.dual-d2.tau-target.mean\endcsname{6.2}
\expandafter\def\csname ppt@rev.desc.d1.sspro.p2.C.dual-d2.tau-target.mean\endcsname{$+$6.2}
\expandafter\def\csname val@rev.desc.d1.sspro.p2.C.dual-d2.tau-target.n-apps\endcsname{6}
\expandafter\def\csname cnt@rev.desc.d1.sspro.p2.C.dual-d2.tau-target.n-apps\endcsname{6}
\expandafter\def\csname pct@rev.desc.d1.sspro.p2.C.dual-d2.tau-target.n-apps\endcsname{600.0}
\expandafter\def\csname ppt@rev.desc.d1.sspro.p2.C.dual-d2.tau-target.n-apps\endcsname{$+$600.0}
\expandafter\def\csname val@rev.desc.d1.sspro.p2.C.dual-d2.tau-target.tost1.ci90-inside\endcsname{0}
\expandafter\def\csname cnt@rev.desc.d1.sspro.p2.C.dual-d2.tau-target.tost1.ci90-inside\endcsname{0}
\expandafter\def\csname pct@rev.desc.d1.sspro.p2.C.dual-d2.tau-target.tost1.ci90-inside\endcsname{0.0}
\expandafter\def\csname ppt@rev.desc.d1.sspro.p2.C.dual-d2.tau-target.tost1.ci90-inside\endcsname{0.0}
\expandafter\def\csname val@rev.desc.d1.sspro.p2.C.dual-d2.tau-target.tost1.equivalent\endcsname{0}
\expandafter\def\csname cnt@rev.desc.d1.sspro.p2.C.dual-d2.tau-target.tost1.equivalent\endcsname{0}
\expandafter\def\csname pct@rev.desc.d1.sspro.p2.C.dual-d2.tau-target.tost1.equivalent\endcsname{0.0}
\expandafter\def\csname ppt@rev.desc.d1.sspro.p2.C.dual-d2.tau-target.tost1.equivalent\endcsname{0.0}
\expandafter\def\csname val@rev.desc.d1.sspro.p2.C.dual-d2.tau-target.tost1.p\endcsname{0.851}
\expandafter\def\csname cnt@rev.desc.d1.sspro.p2.C.dual-d2.tau-target.tost1.p\endcsname{1}
\expandafter\def\csname pct@rev.desc.d1.sspro.p2.C.dual-d2.tau-target.tost1.p\endcsname{85.1}
\expandafter\def\csname ppt@rev.desc.d1.sspro.p2.C.dual-d2.tau-target.tost1.p\endcsname{$+$85.1}
\expandafter\def\csname val@rev.desc.d1.sspro.p2.C.dual-d2.tau-target.tost5.ci90-inside\endcsname{0}
\expandafter\def\csname cnt@rev.desc.d1.sspro.p2.C.dual-d2.tau-target.tost5.ci90-inside\endcsname{0}
\expandafter\def\csname pct@rev.desc.d1.sspro.p2.C.dual-d2.tau-target.tost5.ci90-inside\endcsname{0.0}
\expandafter\def\csname ppt@rev.desc.d1.sspro.p2.C.dual-d2.tau-target.tost5.ci90-inside\endcsname{0.0}
\expandafter\def\csname val@rev.desc.d1.sspro.p2.C.dual-d2.tau-target.tost5.equivalent\endcsname{0}
\expandafter\def\csname cnt@rev.desc.d1.sspro.p2.C.dual-d2.tau-target.tost5.equivalent\endcsname{0}
\expandafter\def\csname pct@rev.desc.d1.sspro.p2.C.dual-d2.tau-target.tost5.equivalent\endcsname{0.0}
\expandafter\def\csname ppt@rev.desc.d1.sspro.p2.C.dual-d2.tau-target.tost5.equivalent\endcsname{0.0}
\expandafter\def\csname val@rev.desc.d1.sspro.p2.C.dual-d2.tau-target.tost5.p\endcsname{0.575}
\expandafter\def\csname cnt@rev.desc.d1.sspro.p2.C.dual-d2.tau-target.tost5.p\endcsname{1}
\expandafter\def\csname pct@rev.desc.d1.sspro.p2.C.dual-d2.tau-target.tost5.p\endcsname{57.5}
\expandafter\def\csname ppt@rev.desc.d1.sspro.p2.C.dual-d2.tau-target.tost5.p\endcsname{$+$57.5}
\expandafter\def\csname val@rev.desc.d1.sspro.p2.C.dual-d2.zero-estimate\endcsname{0}
\expandafter\def\csname cnt@rev.desc.d1.sspro.p2.C.dual-d2.zero-estimate\endcsname{0}
\expandafter\def\csname pct@rev.desc.d1.sspro.p2.C.dual-d2.zero-estimate\endcsname{0.0}
\expandafter\def\csname ppt@rev.desc.d1.sspro.p2.C.dual-d2.zero-estimate\endcsname{0.0}
\expandafter\def\csname val@rev.desc.d1.sspro.p2.C.grid2-d3.delta-tau.ci90.0\endcsname{-0.167}
\expandafter\def\csname cnt@rev.desc.d1.sspro.p2.C.grid2-d3.delta-tau.ci90.0\endcsname{0}
\expandafter\def\csname pct@rev.desc.d1.sspro.p2.C.grid2-d3.delta-tau.ci90.0\endcsname{$-$16.7}
\expandafter\def\csname ppt@rev.desc.d1.sspro.p2.C.grid2-d3.delta-tau.ci90.0\endcsname{$-$16.7}
\expandafter\def\csname val@rev.desc.d1.sspro.p2.C.grid2-d3.delta-tau.ci90.1\endcsname{0.042}
\expandafter\def\csname cnt@rev.desc.d1.sspro.p2.C.grid2-d3.delta-tau.ci90.1\endcsname{0}
\expandafter\def\csname pct@rev.desc.d1.sspro.p2.C.grid2-d3.delta-tau.ci90.1\endcsname{4.2}
\expandafter\def\csname ppt@rev.desc.d1.sspro.p2.C.grid2-d3.delta-tau.ci90.1\endcsname{$+$4.2}
\expandafter\def\csname val@rev.desc.d1.sspro.p2.C.grid2-d3.delta-tau.ci95-excludes-0\endcsname{0}
\expandafter\def\csname cnt@rev.desc.d1.sspro.p2.C.grid2-d3.delta-tau.ci95-excludes-0\endcsname{0}
\expandafter\def\csname pct@rev.desc.d1.sspro.p2.C.grid2-d3.delta-tau.ci95-excludes-0\endcsname{0.0}
\expandafter\def\csname ppt@rev.desc.d1.sspro.p2.C.grid2-d3.delta-tau.ci95-excludes-0\endcsname{0.0}
\expandafter\def\csname val@rev.desc.d1.sspro.p2.C.grid2-d3.delta-tau.ci95.0\endcsname{-0.188}
\expandafter\def\csname cnt@rev.desc.d1.sspro.p2.C.grid2-d3.delta-tau.ci95.0\endcsname{0}
\expandafter\def\csname pct@rev.desc.d1.sspro.p2.C.grid2-d3.delta-tau.ci95.0\endcsname{$-$18.8}
\expandafter\def\csname ppt@rev.desc.d1.sspro.p2.C.grid2-d3.delta-tau.ci95.0\endcsname{$-$18.8}
\expandafter\def\csname val@rev.desc.d1.sspro.p2.C.grid2-d3.delta-tau.ci95.1\endcsname{0.062}
\expandafter\def\csname cnt@rev.desc.d1.sspro.p2.C.grid2-d3.delta-tau.ci95.1\endcsname{0}
\expandafter\def\csname pct@rev.desc.d1.sspro.p2.C.grid2-d3.delta-tau.ci95.1\endcsname{6.2}
\expandafter\def\csname ppt@rev.desc.d1.sspro.p2.C.grid2-d3.delta-tau.ci95.1\endcsname{$+$6.2}
\expandafter\def\csname val@rev.desc.d1.sspro.p2.C.grid2-d3.delta-tau.mean\endcsname{-0.042}
\expandafter\def\csname cnt@rev.desc.d1.sspro.p2.C.grid2-d3.delta-tau.mean\endcsname{0}
\expandafter\def\csname pct@rev.desc.d1.sspro.p2.C.grid2-d3.delta-tau.mean\endcsname{$-$4.2}
\expandafter\def\csname ppt@rev.desc.d1.sspro.p2.C.grid2-d3.delta-tau.mean\endcsname{$-$4.2}
\expandafter\def\csname val@rev.desc.d1.sspro.p2.C.grid2-d3.delta-tau.n-apps\endcsname{6}
\expandafter\def\csname cnt@rev.desc.d1.sspro.p2.C.grid2-d3.delta-tau.n-apps\endcsname{6}
\expandafter\def\csname pct@rev.desc.d1.sspro.p2.C.grid2-d3.delta-tau.n-apps\endcsname{600.0}
\expandafter\def\csname ppt@rev.desc.d1.sspro.p2.C.grid2-d3.delta-tau.n-apps\endcsname{$+$600.0}
\expandafter\def\csname val@rev.desc.d1.sspro.p2.C.grid2-d3.delta-tau.tost1.ci90-inside\endcsname{0}
\expandafter\def\csname cnt@rev.desc.d1.sspro.p2.C.grid2-d3.delta-tau.tost1.ci90-inside\endcsname{0}
\expandafter\def\csname pct@rev.desc.d1.sspro.p2.C.grid2-d3.delta-tau.tost1.ci90-inside\endcsname{0.0}
\expandafter\def\csname ppt@rev.desc.d1.sspro.p2.C.grid2-d3.delta-tau.tost1.ci90-inside\endcsname{0.0}
\expandafter\def\csname val@rev.desc.d1.sspro.p2.C.grid2-d3.delta-tau.tost1.equivalent\endcsname{0}
\expandafter\def\csname cnt@rev.desc.d1.sspro.p2.C.grid2-d3.delta-tau.tost1.equivalent\endcsname{0}
\expandafter\def\csname pct@rev.desc.d1.sspro.p2.C.grid2-d3.delta-tau.tost1.equivalent\endcsname{0.0}
\expandafter\def\csname ppt@rev.desc.d1.sspro.p2.C.grid2-d3.delta-tau.tost1.equivalent\endcsname{0.0}
\expandafter\def\csname val@rev.desc.d1.sspro.p2.C.grid2-d3.delta-tau.tost1.p\endcsname{0.661}
\expandafter\def\csname cnt@rev.desc.d1.sspro.p2.C.grid2-d3.delta-tau.tost1.p\endcsname{1}
\expandafter\def\csname pct@rev.desc.d1.sspro.p2.C.grid2-d3.delta-tau.tost1.p\endcsname{66.1}
\expandafter\def\csname ppt@rev.desc.d1.sspro.p2.C.grid2-d3.delta-tau.tost1.p\endcsname{$+$66.1}
\expandafter\def\csname val@rev.desc.d1.sspro.p2.C.grid2-d3.delta-tau.tost5.ci90-inside\endcsname{0}
\expandafter\def\csname cnt@rev.desc.d1.sspro.p2.C.grid2-d3.delta-tau.tost5.ci90-inside\endcsname{0}
\expandafter\def\csname pct@rev.desc.d1.sspro.p2.C.grid2-d3.delta-tau.tost5.ci90-inside\endcsname{0.0}
\expandafter\def\csname ppt@rev.desc.d1.sspro.p2.C.grid2-d3.delta-tau.tost5.ci90-inside\endcsname{0.0}
\expandafter\def\csname val@rev.desc.d1.sspro.p2.C.grid2-d3.delta-tau.tost5.equivalent\endcsname{0}
\expandafter\def\csname cnt@rev.desc.d1.sspro.p2.C.grid2-d3.delta-tau.tost5.equivalent\endcsname{0}
\expandafter\def\csname pct@rev.desc.d1.sspro.p2.C.grid2-d3.delta-tau.tost5.equivalent\endcsname{0.0}
\expandafter\def\csname ppt@rev.desc.d1.sspro.p2.C.grid2-d3.delta-tau.tost5.equivalent\endcsname{0.0}
\expandafter\def\csname val@rev.desc.d1.sspro.p2.C.grid2-d3.delta-tau.tost5.p\endcsname{0.411}
\expandafter\def\csname cnt@rev.desc.d1.sspro.p2.C.grid2-d3.delta-tau.tost5.p\endcsname{0}
\expandafter\def\csname pct@rev.desc.d1.sspro.p2.C.grid2-d3.delta-tau.tost5.p\endcsname{41.1}
\expandafter\def\csname ppt@rev.desc.d1.sspro.p2.C.grid2-d3.delta-tau.tost5.p\endcsname{$+$41.1}
\expandafter\def\csname val@rev.desc.d1.sspro.p2.C.grid2-d3.material-flip\endcsname{0}
\expandafter\def\csname cnt@rev.desc.d1.sspro.p2.C.grid2-d3.material-flip\endcsname{0}
\expandafter\def\csname pct@rev.desc.d1.sspro.p2.C.grid2-d3.material-flip\endcsname{0.0}
\expandafter\def\csname ppt@rev.desc.d1.sspro.p2.C.grid2-d3.material-flip\endcsname{0.0}
\expandafter\def\csname val@rev.desc.d1.sspro.p2.C.grid2-d3.n-apps\endcsname{6}
\expandafter\def\csname cnt@rev.desc.d1.sspro.p2.C.grid2-d3.n-apps\endcsname{6}
\expandafter\def\csname pct@rev.desc.d1.sspro.p2.C.grid2-d3.n-apps\endcsname{600.0}
\expandafter\def\csname ppt@rev.desc.d1.sspro.p2.C.grid2-d3.n-apps\endcsname{$+$600.0}
\expandafter\def\csname val@rev.desc.d1.sspro.p2.C.grid2-d3.n-tasks\endcsname{48}
\expandafter\def\csname cnt@rev.desc.d1.sspro.p2.C.grid2-d3.n-tasks\endcsname{48}
\expandafter\def\csname pct@rev.desc.d1.sspro.p2.C.grid2-d3.n-tasks\endcsname{4800.0}
\expandafter\def\csname ppt@rev.desc.d1.sspro.p2.C.grid2-d3.n-tasks\endcsname{$+$4800.0}
\expandafter\def\csname val@rev.desc.d1.sspro.p2.C.grid2-d3.sign-differs\endcsname{0}
\expandafter\def\csname cnt@rev.desc.d1.sspro.p2.C.grid2-d3.sign-differs\endcsname{0}
\expandafter\def\csname pct@rev.desc.d1.sspro.p2.C.grid2-d3.sign-differs\endcsname{0.0}
\expandafter\def\csname ppt@rev.desc.d1.sspro.p2.C.grid2-d3.sign-differs\endcsname{0.0}
\expandafter\def\csname val@rev.desc.d1.sspro.p2.C.grid2-d3.tau-source.ci90.0\endcsname{0.062}
\expandafter\def\csname cnt@rev.desc.d1.sspro.p2.C.grid2-d3.tau-source.ci90.0\endcsname{0}
\expandafter\def\csname pct@rev.desc.d1.sspro.p2.C.grid2-d3.tau-source.ci90.0\endcsname{6.2}
\expandafter\def\csname ppt@rev.desc.d1.sspro.p2.C.grid2-d3.tau-source.ci90.0\endcsname{$+$6.2}
\expandafter\def\csname val@rev.desc.d1.sspro.p2.C.grid2-d3.tau-source.ci90.1\endcsname{0.250}
\expandafter\def\csname cnt@rev.desc.d1.sspro.p2.C.grid2-d3.tau-source.ci90.1\endcsname{0}
\expandafter\def\csname pct@rev.desc.d1.sspro.p2.C.grid2-d3.tau-source.ci90.1\endcsname{25.0}
\expandafter\def\csname ppt@rev.desc.d1.sspro.p2.C.grid2-d3.tau-source.ci90.1\endcsname{$+$25.0}
\expandafter\def\csname val@rev.desc.d1.sspro.p2.C.grid2-d3.tau-source.ci95-excludes-0\endcsname{1}
\expandafter\def\csname cnt@rev.desc.d1.sspro.p2.C.grid2-d3.tau-source.ci95-excludes-0\endcsname{1}
\expandafter\def\csname pct@rev.desc.d1.sspro.p2.C.grid2-d3.tau-source.ci95-excludes-0\endcsname{100.0}
\expandafter\def\csname ppt@rev.desc.d1.sspro.p2.C.grid2-d3.tau-source.ci95-excludes-0\endcsname{$+$100.0}
\expandafter\def\csname val@rev.desc.d1.sspro.p2.C.grid2-d3.tau-source.ci95.0\endcsname{0.042}
\expandafter\def\csname cnt@rev.desc.d1.sspro.p2.C.grid2-d3.tau-source.ci95.0\endcsname{0}
\expandafter\def\csname pct@rev.desc.d1.sspro.p2.C.grid2-d3.tau-source.ci95.0\endcsname{4.2}
\expandafter\def\csname ppt@rev.desc.d1.sspro.p2.C.grid2-d3.tau-source.ci95.0\endcsname{$+$4.2}
\expandafter\def\csname val@rev.desc.d1.sspro.p2.C.grid2-d3.tau-source.ci95.1\endcsname{0.250}
\expandafter\def\csname cnt@rev.desc.d1.sspro.p2.C.grid2-d3.tau-source.ci95.1\endcsname{0}
\expandafter\def\csname pct@rev.desc.d1.sspro.p2.C.grid2-d3.tau-source.ci95.1\endcsname{25.0}
\expandafter\def\csname ppt@rev.desc.d1.sspro.p2.C.grid2-d3.tau-source.ci95.1\endcsname{$+$25.0}
\expandafter\def\csname val@rev.desc.d1.sspro.p2.C.grid2-d3.tau-source.mean\endcsname{0.146}
\expandafter\def\csname cnt@rev.desc.d1.sspro.p2.C.grid2-d3.tau-source.mean\endcsname{0}
\expandafter\def\csname pct@rev.desc.d1.sspro.p2.C.grid2-d3.tau-source.mean\endcsname{14.6}
\expandafter\def\csname ppt@rev.desc.d1.sspro.p2.C.grid2-d3.tau-source.mean\endcsname{$+$14.6}
\expandafter\def\csname val@rev.desc.d1.sspro.p2.C.grid2-d3.tau-source.n-apps\endcsname{6}
\expandafter\def\csname cnt@rev.desc.d1.sspro.p2.C.grid2-d3.tau-source.n-apps\endcsname{6}
\expandafter\def\csname pct@rev.desc.d1.sspro.p2.C.grid2-d3.tau-source.n-apps\endcsname{600.0}
\expandafter\def\csname ppt@rev.desc.d1.sspro.p2.C.grid2-d3.tau-source.n-apps\endcsname{$+$600.0}
\expandafter\def\csname val@rev.desc.d1.sspro.p2.C.grid2-d3.tau-source.tost1.ci90-inside\endcsname{0}
\expandafter\def\csname cnt@rev.desc.d1.sspro.p2.C.grid2-d3.tau-source.tost1.ci90-inside\endcsname{0}
\expandafter\def\csname pct@rev.desc.d1.sspro.p2.C.grid2-d3.tau-source.tost1.ci90-inside\endcsname{0.0}
\expandafter\def\csname ppt@rev.desc.d1.sspro.p2.C.grid2-d3.tau-source.tost1.ci90-inside\endcsname{0.0}
\expandafter\def\csname val@rev.desc.d1.sspro.p2.C.grid2-d3.tau-source.tost1.equivalent\endcsname{0}
\expandafter\def\csname cnt@rev.desc.d1.sspro.p2.C.grid2-d3.tau-source.tost1.equivalent\endcsname{0}
\expandafter\def\csname pct@rev.desc.d1.sspro.p2.C.grid2-d3.tau-source.tost1.equivalent\endcsname{0.0}
\expandafter\def\csname ppt@rev.desc.d1.sspro.p2.C.grid2-d3.tau-source.tost1.equivalent\endcsname{0.0}
\expandafter\def\csname val@rev.desc.d1.sspro.p2.C.grid2-d3.tau-source.tost1.p\endcsname{0.999}
\expandafter\def\csname cnt@rev.desc.d1.sspro.p2.C.grid2-d3.tau-source.tost1.p\endcsname{1}
\expandafter\def\csname pct@rev.desc.d1.sspro.p2.C.grid2-d3.tau-source.tost1.p\endcsname{99.9}
\expandafter\def\csname ppt@rev.desc.d1.sspro.p2.C.grid2-d3.tau-source.tost1.p\endcsname{$+$99.9}
\expandafter\def\csname val@rev.desc.d1.sspro.p2.C.grid2-d3.tau-source.tost5.ci90-inside\endcsname{0}
\expandafter\def\csname cnt@rev.desc.d1.sspro.p2.C.grid2-d3.tau-source.tost5.ci90-inside\endcsname{0}
\expandafter\def\csname pct@rev.desc.d1.sspro.p2.C.grid2-d3.tau-source.tost5.ci90-inside\endcsname{0.0}
\expandafter\def\csname ppt@rev.desc.d1.sspro.p2.C.grid2-d3.tau-source.tost5.ci90-inside\endcsname{0.0}
\expandafter\def\csname val@rev.desc.d1.sspro.p2.C.grid2-d3.tau-source.tost5.equivalent\endcsname{0}
\expandafter\def\csname cnt@rev.desc.d1.sspro.p2.C.grid2-d3.tau-source.tost5.equivalent\endcsname{0}
\expandafter\def\csname pct@rev.desc.d1.sspro.p2.C.grid2-d3.tau-source.tost5.equivalent\endcsname{0.0}
\expandafter\def\csname ppt@rev.desc.d1.sspro.p2.C.grid2-d3.tau-source.tost5.equivalent\endcsname{0.0}
\expandafter\def\csname val@rev.desc.d1.sspro.p2.C.grid2-d3.tau-source.tost5.p\endcsname{0.963}
\expandafter\def\csname cnt@rev.desc.d1.sspro.p2.C.grid2-d3.tau-source.tost5.p\endcsname{1}
\expandafter\def\csname pct@rev.desc.d1.sspro.p2.C.grid2-d3.tau-source.tost5.p\endcsname{96.3}
\expandafter\def\csname ppt@rev.desc.d1.sspro.p2.C.grid2-d3.tau-source.tost5.p\endcsname{$+$96.3}
\expandafter\def\csname val@rev.desc.d1.sspro.p2.C.grid2-d3.tau-target.ci90.0\endcsname{0.042}
\expandafter\def\csname cnt@rev.desc.d1.sspro.p2.C.grid2-d3.tau-target.ci90.0\endcsname{0}
\expandafter\def\csname pct@rev.desc.d1.sspro.p2.C.grid2-d3.tau-target.ci90.0\endcsname{4.2}
\expandafter\def\csname ppt@rev.desc.d1.sspro.p2.C.grid2-d3.tau-target.ci90.0\endcsname{$+$4.2}
\expandafter\def\csname val@rev.desc.d1.sspro.p2.C.grid2-d3.tau-target.ci90.1\endcsname{0.167}
\expandafter\def\csname cnt@rev.desc.d1.sspro.p2.C.grid2-d3.tau-target.ci90.1\endcsname{0}
\expandafter\def\csname pct@rev.desc.d1.sspro.p2.C.grid2-d3.tau-target.ci90.1\endcsname{16.7}
\expandafter\def\csname ppt@rev.desc.d1.sspro.p2.C.grid2-d3.tau-target.ci90.1\endcsname{$+$16.7}
\expandafter\def\csname val@rev.desc.d1.sspro.p2.C.grid2-d3.tau-target.ci95-excludes-0\endcsname{1}
\expandafter\def\csname cnt@rev.desc.d1.sspro.p2.C.grid2-d3.tau-target.ci95-excludes-0\endcsname{1}
\expandafter\def\csname pct@rev.desc.d1.sspro.p2.C.grid2-d3.tau-target.ci95-excludes-0\endcsname{100.0}
\expandafter\def\csname ppt@rev.desc.d1.sspro.p2.C.grid2-d3.tau-target.ci95-excludes-0\endcsname{$+$100.0}
\expandafter\def\csname val@rev.desc.d1.sspro.p2.C.grid2-d3.tau-target.ci95.0\endcsname{0.042}
\expandafter\def\csname cnt@rev.desc.d1.sspro.p2.C.grid2-d3.tau-target.ci95.0\endcsname{0}
\expandafter\def\csname pct@rev.desc.d1.sspro.p2.C.grid2-d3.tau-target.ci95.0\endcsname{4.2}
\expandafter\def\csname ppt@rev.desc.d1.sspro.p2.C.grid2-d3.tau-target.ci95.0\endcsname{$+$4.2}
\expandafter\def\csname val@rev.desc.d1.sspro.p2.C.grid2-d3.tau-target.ci95.1\endcsname{0.167}
\expandafter\def\csname cnt@rev.desc.d1.sspro.p2.C.grid2-d3.tau-target.ci95.1\endcsname{0}
\expandafter\def\csname pct@rev.desc.d1.sspro.p2.C.grid2-d3.tau-target.ci95.1\endcsname{16.7}
\expandafter\def\csname ppt@rev.desc.d1.sspro.p2.C.grid2-d3.tau-target.ci95.1\endcsname{$+$16.7}
\expandafter\def\csname val@rev.desc.d1.sspro.p2.C.grid2-d3.tau-target.mean\endcsname{0.104}
\expandafter\def\csname cnt@rev.desc.d1.sspro.p2.C.grid2-d3.tau-target.mean\endcsname{0}
\expandafter\def\csname pct@rev.desc.d1.sspro.p2.C.grid2-d3.tau-target.mean\endcsname{10.4}
\expandafter\def\csname ppt@rev.desc.d1.sspro.p2.C.grid2-d3.tau-target.mean\endcsname{$+$10.4}
\expandafter\def\csname val@rev.desc.d1.sspro.p2.C.grid2-d3.tau-target.n-apps\endcsname{6}
\expandafter\def\csname cnt@rev.desc.d1.sspro.p2.C.grid2-d3.tau-target.n-apps\endcsname{6}
\expandafter\def\csname pct@rev.desc.d1.sspro.p2.C.grid2-d3.tau-target.n-apps\endcsname{600.0}
\expandafter\def\csname ppt@rev.desc.d1.sspro.p2.C.grid2-d3.tau-target.n-apps\endcsname{$+$600.0}
\expandafter\def\csname val@rev.desc.d1.sspro.p2.C.grid2-d3.tau-target.tost1.ci90-inside\endcsname{0}
\expandafter\def\csname cnt@rev.desc.d1.sspro.p2.C.grid2-d3.tau-target.tost1.ci90-inside\endcsname{0}
\expandafter\def\csname pct@rev.desc.d1.sspro.p2.C.grid2-d3.tau-target.tost1.ci90-inside\endcsname{0.0}
\expandafter\def\csname ppt@rev.desc.d1.sspro.p2.C.grid2-d3.tau-target.tost1.ci90-inside\endcsname{0.0}
\expandafter\def\csname val@rev.desc.d1.sspro.p2.C.grid2-d3.tau-target.tost1.equivalent\endcsname{0}
\expandafter\def\csname cnt@rev.desc.d1.sspro.p2.C.grid2-d3.tau-target.tost1.equivalent\endcsname{0}
\expandafter\def\csname pct@rev.desc.d1.sspro.p2.C.grid2-d3.tau-target.tost1.equivalent\endcsname{0.0}
\expandafter\def\csname ppt@rev.desc.d1.sspro.p2.C.grid2-d3.tau-target.tost1.equivalent\endcsname{0.0}
\expandafter\def\csname val@rev.desc.d1.sspro.p2.C.grid2-d3.tau-target.tost1.p\endcsname{0.999}
\expandafter\def\csname cnt@rev.desc.d1.sspro.p2.C.grid2-d3.tau-target.tost1.p\endcsname{1}
\expandafter\def\csname pct@rev.desc.d1.sspro.p2.C.grid2-d3.tau-target.tost1.p\endcsname{99.9}
\expandafter\def\csname ppt@rev.desc.d1.sspro.p2.C.grid2-d3.tau-target.tost1.p\endcsname{$+$99.9}
\expandafter\def\csname val@rev.desc.d1.sspro.p2.C.grid2-d3.tau-target.tost5.ci90-inside\endcsname{0}
\expandafter\def\csname cnt@rev.desc.d1.sspro.p2.C.grid2-d3.tau-target.tost5.ci90-inside\endcsname{0}
\expandafter\def\csname pct@rev.desc.d1.sspro.p2.C.grid2-d3.tau-target.tost5.ci90-inside\endcsname{0.0}
\expandafter\def\csname ppt@rev.desc.d1.sspro.p2.C.grid2-d3.tau-target.tost5.ci90-inside\endcsname{0.0}
\expandafter\def\csname val@rev.desc.d1.sspro.p2.C.grid2-d3.tau-target.tost5.equivalent\endcsname{0}
\expandafter\def\csname cnt@rev.desc.d1.sspro.p2.C.grid2-d3.tau-target.tost5.equivalent\endcsname{0}
\expandafter\def\csname pct@rev.desc.d1.sspro.p2.C.grid2-d3.tau-target.tost5.equivalent\endcsname{0.0}
\expandafter\def\csname ppt@rev.desc.d1.sspro.p2.C.grid2-d3.tau-target.tost5.equivalent\endcsname{0.0}
\expandafter\def\csname val@rev.desc.d1.sspro.p2.C.grid2-d3.tau-target.tost5.p\endcsname{0.942}
\expandafter\def\csname cnt@rev.desc.d1.sspro.p2.C.grid2-d3.tau-target.tost5.p\endcsname{1}
\expandafter\def\csname pct@rev.desc.d1.sspro.p2.C.grid2-d3.tau-target.tost5.p\endcsname{94.2}
\expandafter\def\csname ppt@rev.desc.d1.sspro.p2.C.grid2-d3.tau-target.tost5.p\endcsname{$+$94.2}
\expandafter\def\csname val@rev.desc.d1.sspro.p2.C.grid2-d3.zero-estimate\endcsname{0}
\expandafter\def\csname cnt@rev.desc.d1.sspro.p2.C.grid2-d3.zero-estimate\endcsname{0}
\expandafter\def\csname pct@rev.desc.d1.sspro.p2.C.grid2-d3.zero-estimate\endcsname{0.0}
\expandafter\def\csname ppt@rev.desc.d1.sspro.p2.C.grid2-d3.zero-estimate\endcsname{0.0}
\expandafter\def\csname val@rev.desc.d1.sspro.p2.C.zoomclick-d3.delta-tau.ci90.0\endcsname{-0.146}
\expandafter\def\csname cnt@rev.desc.d1.sspro.p2.C.zoomclick-d3.delta-tau.ci90.0\endcsname{0}
\expandafter\def\csname pct@rev.desc.d1.sspro.p2.C.zoomclick-d3.delta-tau.ci90.0\endcsname{$-$14.6}
\expandafter\def\csname ppt@rev.desc.d1.sspro.p2.C.zoomclick-d3.delta-tau.ci90.0\endcsname{$-$14.6}
\expandafter\def\csname val@rev.desc.d1.sspro.p2.C.zoomclick-d3.delta-tau.ci90.1\endcsname{0.062}
\expandafter\def\csname cnt@rev.desc.d1.sspro.p2.C.zoomclick-d3.delta-tau.ci90.1\endcsname{0}
\expandafter\def\csname pct@rev.desc.d1.sspro.p2.C.zoomclick-d3.delta-tau.ci90.1\endcsname{6.2}
\expandafter\def\csname ppt@rev.desc.d1.sspro.p2.C.zoomclick-d3.delta-tau.ci90.1\endcsname{$+$6.2}
\expandafter\def\csname val@rev.desc.d1.sspro.p2.C.zoomclick-d3.delta-tau.ci95-excludes-0\endcsname{0}
\expandafter\def\csname cnt@rev.desc.d1.sspro.p2.C.zoomclick-d3.delta-tau.ci95-excludes-0\endcsname{0}
\expandafter\def\csname pct@rev.desc.d1.sspro.p2.C.zoomclick-d3.delta-tau.ci95-excludes-0\endcsname{0.0}
\expandafter\def\csname ppt@rev.desc.d1.sspro.p2.C.zoomclick-d3.delta-tau.ci95-excludes-0\endcsname{0.0}
\expandafter\def\csname val@rev.desc.d1.sspro.p2.C.zoomclick-d3.delta-tau.ci95.0\endcsname{-0.167}
\expandafter\def\csname cnt@rev.desc.d1.sspro.p2.C.zoomclick-d3.delta-tau.ci95.0\endcsname{0}
\expandafter\def\csname pct@rev.desc.d1.sspro.p2.C.zoomclick-d3.delta-tau.ci95.0\endcsname{$-$16.7}
\expandafter\def\csname ppt@rev.desc.d1.sspro.p2.C.zoomclick-d3.delta-tau.ci95.0\endcsname{$-$16.7}
\expandafter\def\csname val@rev.desc.d1.sspro.p2.C.zoomclick-d3.delta-tau.ci95.1\endcsname{0.083}
\expandafter\def\csname cnt@rev.desc.d1.sspro.p2.C.zoomclick-d3.delta-tau.ci95.1\endcsname{0}
\expandafter\def\csname pct@rev.desc.d1.sspro.p2.C.zoomclick-d3.delta-tau.ci95.1\endcsname{8.3}
\expandafter\def\csname ppt@rev.desc.d1.sspro.p2.C.zoomclick-d3.delta-tau.ci95.1\endcsname{$+$8.3}
\expandafter\def\csname val@rev.desc.d1.sspro.p2.C.zoomclick-d3.delta-tau.mean\endcsname{-0.042}
\expandafter\def\csname cnt@rev.desc.d1.sspro.p2.C.zoomclick-d3.delta-tau.mean\endcsname{0}
\expandafter\def\csname pct@rev.desc.d1.sspro.p2.C.zoomclick-d3.delta-tau.mean\endcsname{$-$4.2}
\expandafter\def\csname ppt@rev.desc.d1.sspro.p2.C.zoomclick-d3.delta-tau.mean\endcsname{$-$4.2}
\expandafter\def\csname val@rev.desc.d1.sspro.p2.C.zoomclick-d3.delta-tau.n-apps\endcsname{6}
\expandafter\def\csname cnt@rev.desc.d1.sspro.p2.C.zoomclick-d3.delta-tau.n-apps\endcsname{6}
\expandafter\def\csname pct@rev.desc.d1.sspro.p2.C.zoomclick-d3.delta-tau.n-apps\endcsname{600.0}
\expandafter\def\csname ppt@rev.desc.d1.sspro.p2.C.zoomclick-d3.delta-tau.n-apps\endcsname{$+$600.0}
\expandafter\def\csname val@rev.desc.d1.sspro.p2.C.zoomclick-d3.delta-tau.tost1.ci90-inside\endcsname{0}
\expandafter\def\csname cnt@rev.desc.d1.sspro.p2.C.zoomclick-d3.delta-tau.tost1.ci90-inside\endcsname{0}
\expandafter\def\csname pct@rev.desc.d1.sspro.p2.C.zoomclick-d3.delta-tau.tost1.ci90-inside\endcsname{0.0}
\expandafter\def\csname ppt@rev.desc.d1.sspro.p2.C.zoomclick-d3.delta-tau.tost1.ci90-inside\endcsname{0.0}
\expandafter\def\csname val@rev.desc.d1.sspro.p2.C.zoomclick-d3.delta-tau.tost1.equivalent\endcsname{0}
\expandafter\def\csname cnt@rev.desc.d1.sspro.p2.C.zoomclick-d3.delta-tau.tost1.equivalent\endcsname{0}
\expandafter\def\csname pct@rev.desc.d1.sspro.p2.C.zoomclick-d3.delta-tau.tost1.equivalent\endcsname{0.0}
\expandafter\def\csname ppt@rev.desc.d1.sspro.p2.C.zoomclick-d3.delta-tau.tost1.equivalent\endcsname{0.0}
\expandafter\def\csname val@rev.desc.d1.sspro.p2.C.zoomclick-d3.delta-tau.tost1.p\endcsname{0.701}
\expandafter\def\csname cnt@rev.desc.d1.sspro.p2.C.zoomclick-d3.delta-tau.tost1.p\endcsname{1}
\expandafter\def\csname pct@rev.desc.d1.sspro.p2.C.zoomclick-d3.delta-tau.tost1.p\endcsname{70.1}
\expandafter\def\csname ppt@rev.desc.d1.sspro.p2.C.zoomclick-d3.delta-tau.tost1.p\endcsname{$+$70.1}
\expandafter\def\csname val@rev.desc.d1.sspro.p2.C.zoomclick-d3.delta-tau.tost5.ci90-inside\endcsname{0}
\expandafter\def\csname cnt@rev.desc.d1.sspro.p2.C.zoomclick-d3.delta-tau.tost5.ci90-inside\endcsname{0}
\expandafter\def\csname pct@rev.desc.d1.sspro.p2.C.zoomclick-d3.delta-tau.tost5.ci90-inside\endcsname{0.0}
\expandafter\def\csname ppt@rev.desc.d1.sspro.p2.C.zoomclick-d3.delta-tau.tost5.ci90-inside\endcsname{0.0}
\expandafter\def\csname val@rev.desc.d1.sspro.p2.C.zoomclick-d3.delta-tau.tost5.equivalent\endcsname{0}
\expandafter\def\csname cnt@rev.desc.d1.sspro.p2.C.zoomclick-d3.delta-tau.tost5.equivalent\endcsname{0}
\expandafter\def\csname pct@rev.desc.d1.sspro.p2.C.zoomclick-d3.delta-tau.tost5.equivalent\endcsname{0.0}
\expandafter\def\csname ppt@rev.desc.d1.sspro.p2.C.zoomclick-d3.delta-tau.tost5.equivalent\endcsname{0.0}
\expandafter\def\csname val@rev.desc.d1.sspro.p2.C.zoomclick-d3.delta-tau.tost5.p\endcsname{0.456}
\expandafter\def\csname cnt@rev.desc.d1.sspro.p2.C.zoomclick-d3.delta-tau.tost5.p\endcsname{0}
\expandafter\def\csname pct@rev.desc.d1.sspro.p2.C.zoomclick-d3.delta-tau.tost5.p\endcsname{45.6}
\expandafter\def\csname ppt@rev.desc.d1.sspro.p2.C.zoomclick-d3.delta-tau.tost5.p\endcsname{$+$45.6}
\expandafter\def\csname val@rev.desc.d1.sspro.p2.C.zoomclick-d3.material-flip\endcsname{0}
\expandafter\def\csname cnt@rev.desc.d1.sspro.p2.C.zoomclick-d3.material-flip\endcsname{0}
\expandafter\def\csname pct@rev.desc.d1.sspro.p2.C.zoomclick-d3.material-flip\endcsname{0.0}
\expandafter\def\csname ppt@rev.desc.d1.sspro.p2.C.zoomclick-d3.material-flip\endcsname{0.0}
\expandafter\def\csname val@rev.desc.d1.sspro.p2.C.zoomclick-d3.n-apps\endcsname{6}
\expandafter\def\csname cnt@rev.desc.d1.sspro.p2.C.zoomclick-d3.n-apps\endcsname{6}
\expandafter\def\csname pct@rev.desc.d1.sspro.p2.C.zoomclick-d3.n-apps\endcsname{600.0}
\expandafter\def\csname ppt@rev.desc.d1.sspro.p2.C.zoomclick-d3.n-apps\endcsname{$+$600.0}
\expandafter\def\csname val@rev.desc.d1.sspro.p2.C.zoomclick-d3.n-tasks\endcsname{48}
\expandafter\def\csname cnt@rev.desc.d1.sspro.p2.C.zoomclick-d3.n-tasks\endcsname{48}
\expandafter\def\csname pct@rev.desc.d1.sspro.p2.C.zoomclick-d3.n-tasks\endcsname{4800.0}
\expandafter\def\csname ppt@rev.desc.d1.sspro.p2.C.zoomclick-d3.n-tasks\endcsname{$+$4800.0}
\expandafter\def\csname val@rev.desc.d1.sspro.p2.C.zoomclick-d3.sign-differs\endcsname{0}
\expandafter\def\csname cnt@rev.desc.d1.sspro.p2.C.zoomclick-d3.sign-differs\endcsname{0}
\expandafter\def\csname pct@rev.desc.d1.sspro.p2.C.zoomclick-d3.sign-differs\endcsname{0.0}
\expandafter\def\csname ppt@rev.desc.d1.sspro.p2.C.zoomclick-d3.sign-differs\endcsname{0.0}
\expandafter\def\csname val@rev.desc.d1.sspro.p2.C.zoomclick-d3.tau-source.ci90.0\endcsname{0.125}
\expandafter\def\csname cnt@rev.desc.d1.sspro.p2.C.zoomclick-d3.tau-source.ci90.0\endcsname{0}
\expandafter\def\csname pct@rev.desc.d1.sspro.p2.C.zoomclick-d3.tau-source.ci90.0\endcsname{12.5}
\expandafter\def\csname ppt@rev.desc.d1.sspro.p2.C.zoomclick-d3.tau-source.ci90.0\endcsname{$+$12.5}
\expandafter\def\csname val@rev.desc.d1.sspro.p2.C.zoomclick-d3.tau-source.ci90.1\endcsname{0.292}
\expandafter\def\csname cnt@rev.desc.d1.sspro.p2.C.zoomclick-d3.tau-source.ci90.1\endcsname{0}
\expandafter\def\csname pct@rev.desc.d1.sspro.p2.C.zoomclick-d3.tau-source.ci90.1\endcsname{29.2}
\expandafter\def\csname ppt@rev.desc.d1.sspro.p2.C.zoomclick-d3.tau-source.ci90.1\endcsname{$+$29.2}
\expandafter\def\csname val@rev.desc.d1.sspro.p2.C.zoomclick-d3.tau-source.ci95-excludes-0\endcsname{1}
\expandafter\def\csname cnt@rev.desc.d1.sspro.p2.C.zoomclick-d3.tau-source.ci95-excludes-0\endcsname{1}
\expandafter\def\csname pct@rev.desc.d1.sspro.p2.C.zoomclick-d3.tau-source.ci95-excludes-0\endcsname{100.0}
\expandafter\def\csname ppt@rev.desc.d1.sspro.p2.C.zoomclick-d3.tau-source.ci95-excludes-0\endcsname{$+$100.0}
\expandafter\def\csname val@rev.desc.d1.sspro.p2.C.zoomclick-d3.tau-source.ci95.0\endcsname{0.125}
\expandafter\def\csname cnt@rev.desc.d1.sspro.p2.C.zoomclick-d3.tau-source.ci95.0\endcsname{0}
\expandafter\def\csname pct@rev.desc.d1.sspro.p2.C.zoomclick-d3.tau-source.ci95.0\endcsname{12.5}
\expandafter\def\csname ppt@rev.desc.d1.sspro.p2.C.zoomclick-d3.tau-source.ci95.0\endcsname{$+$12.5}
\expandafter\def\csname val@rev.desc.d1.sspro.p2.C.zoomclick-d3.tau-source.ci95.1\endcsname{0.292}
\expandafter\def\csname cnt@rev.desc.d1.sspro.p2.C.zoomclick-d3.tau-source.ci95.1\endcsname{0}
\expandafter\def\csname pct@rev.desc.d1.sspro.p2.C.zoomclick-d3.tau-source.ci95.1\endcsname{29.2}
\expandafter\def\csname ppt@rev.desc.d1.sspro.p2.C.zoomclick-d3.tau-source.ci95.1\endcsname{$+$29.2}
\expandafter\def\csname val@rev.desc.d1.sspro.p2.C.zoomclick-d3.tau-source.mean\endcsname{0.208}
\expandafter\def\csname cnt@rev.desc.d1.sspro.p2.C.zoomclick-d3.tau-source.mean\endcsname{0}
\expandafter\def\csname pct@rev.desc.d1.sspro.p2.C.zoomclick-d3.tau-source.mean\endcsname{20.8}
\expandafter\def\csname ppt@rev.desc.d1.sspro.p2.C.zoomclick-d3.tau-source.mean\endcsname{$+$20.8}
\expandafter\def\csname val@rev.desc.d1.sspro.p2.C.zoomclick-d3.tau-source.n-apps\endcsname{6}
\expandafter\def\csname cnt@rev.desc.d1.sspro.p2.C.zoomclick-d3.tau-source.n-apps\endcsname{6}
\expandafter\def\csname pct@rev.desc.d1.sspro.p2.C.zoomclick-d3.tau-source.n-apps\endcsname{600.0}
\expandafter\def\csname ppt@rev.desc.d1.sspro.p2.C.zoomclick-d3.tau-source.n-apps\endcsname{$+$600.0}
\expandafter\def\csname val@rev.desc.d1.sspro.p2.C.zoomclick-d3.tau-source.tost1.ci90-inside\endcsname{0}
\expandafter\def\csname cnt@rev.desc.d1.sspro.p2.C.zoomclick-d3.tau-source.tost1.ci90-inside\endcsname{0}
\expandafter\def\csname pct@rev.desc.d1.sspro.p2.C.zoomclick-d3.tau-source.tost1.ci90-inside\endcsname{0.0}
\expandafter\def\csname ppt@rev.desc.d1.sspro.p2.C.zoomclick-d3.tau-source.tost1.ci90-inside\endcsname{0.0}
\expandafter\def\csname val@rev.desc.d1.sspro.p2.C.zoomclick-d3.tau-source.tost1.equivalent\endcsname{0}
\expandafter\def\csname cnt@rev.desc.d1.sspro.p2.C.zoomclick-d3.tau-source.tost1.equivalent\endcsname{0}
\expandafter\def\csname pct@rev.desc.d1.sspro.p2.C.zoomclick-d3.tau-source.tost1.equivalent\endcsname{0.0}
\expandafter\def\csname ppt@rev.desc.d1.sspro.p2.C.zoomclick-d3.tau-source.tost1.equivalent\endcsname{0.0}
\expandafter\def\csname val@rev.desc.d1.sspro.p2.C.zoomclick-d3.tau-source.tost1.p\endcsname{1.000}
\expandafter\def\csname cnt@rev.desc.d1.sspro.p2.C.zoomclick-d3.tau-source.tost1.p\endcsname{1}
\expandafter\def\csname pct@rev.desc.d1.sspro.p2.C.zoomclick-d3.tau-source.tost1.p\endcsname{100.0}
\expandafter\def\csname ppt@rev.desc.d1.sspro.p2.C.zoomclick-d3.tau-source.tost1.p\endcsname{$+$100.0}
\expandafter\def\csname val@rev.desc.d1.sspro.p2.C.zoomclick-d3.tau-source.tost5.ci90-inside\endcsname{0}
\expandafter\def\csname cnt@rev.desc.d1.sspro.p2.C.zoomclick-d3.tau-source.tost5.ci90-inside\endcsname{0}
\expandafter\def\csname pct@rev.desc.d1.sspro.p2.C.zoomclick-d3.tau-source.tost5.ci90-inside\endcsname{0.0}
\expandafter\def\csname ppt@rev.desc.d1.sspro.p2.C.zoomclick-d3.tau-source.tost5.ci90-inside\endcsname{0.0}
\expandafter\def\csname val@rev.desc.d1.sspro.p2.C.zoomclick-d3.tau-source.tost5.equivalent\endcsname{0}
\expandafter\def\csname cnt@rev.desc.d1.sspro.p2.C.zoomclick-d3.tau-source.tost5.equivalent\endcsname{0}
\expandafter\def\csname pct@rev.desc.d1.sspro.p2.C.zoomclick-d3.tau-source.tost5.equivalent\endcsname{0.0}
\expandafter\def\csname ppt@rev.desc.d1.sspro.p2.C.zoomclick-d3.tau-source.tost5.equivalent\endcsname{0.0}
\expandafter\def\csname val@rev.desc.d1.sspro.p2.C.zoomclick-d3.tau-source.tost5.p\endcsname{1.000}
\expandafter\def\csname cnt@rev.desc.d1.sspro.p2.C.zoomclick-d3.tau-source.tost5.p\endcsname{1}
\expandafter\def\csname pct@rev.desc.d1.sspro.p2.C.zoomclick-d3.tau-source.tost5.p\endcsname{100.0}
\expandafter\def\csname ppt@rev.desc.d1.sspro.p2.C.zoomclick-d3.tau-source.tost5.p\endcsname{$+$100.0}
\expandafter\def\csname val@rev.desc.d1.sspro.p2.C.zoomclick-d3.tau-target.ci90.0\endcsname{0.083}
\expandafter\def\csname cnt@rev.desc.d1.sspro.p2.C.zoomclick-d3.tau-target.ci90.0\endcsname{0}
\expandafter\def\csname pct@rev.desc.d1.sspro.p2.C.zoomclick-d3.tau-target.ci90.0\endcsname{8.3}
\expandafter\def\csname ppt@rev.desc.d1.sspro.p2.C.zoomclick-d3.tau-target.ci90.0\endcsname{$+$8.3}
\expandafter\def\csname val@rev.desc.d1.sspro.p2.C.zoomclick-d3.tau-target.ci90.1\endcsname{0.250}
\expandafter\def\csname cnt@rev.desc.d1.sspro.p2.C.zoomclick-d3.tau-target.ci90.1\endcsname{0}
\expandafter\def\csname pct@rev.desc.d1.sspro.p2.C.zoomclick-d3.tau-target.ci90.1\endcsname{25.0}
\expandafter\def\csname ppt@rev.desc.d1.sspro.p2.C.zoomclick-d3.tau-target.ci90.1\endcsname{$+$25.0}
\expandafter\def\csname val@rev.desc.d1.sspro.p2.C.zoomclick-d3.tau-target.ci95-excludes-0\endcsname{1}
\expandafter\def\csname cnt@rev.desc.d1.sspro.p2.C.zoomclick-d3.tau-target.ci95-excludes-0\endcsname{1}
\expandafter\def\csname pct@rev.desc.d1.sspro.p2.C.zoomclick-d3.tau-target.ci95-excludes-0\endcsname{100.0}
\expandafter\def\csname ppt@rev.desc.d1.sspro.p2.C.zoomclick-d3.tau-target.ci95-excludes-0\endcsname{$+$100.0}
\expandafter\def\csname val@rev.desc.d1.sspro.p2.C.zoomclick-d3.tau-target.ci95.0\endcsname{0.083}
\expandafter\def\csname cnt@rev.desc.d1.sspro.p2.C.zoomclick-d3.tau-target.ci95.0\endcsname{0}
\expandafter\def\csname pct@rev.desc.d1.sspro.p2.C.zoomclick-d3.tau-target.ci95.0\endcsname{8.3}
\expandafter\def\csname ppt@rev.desc.d1.sspro.p2.C.zoomclick-d3.tau-target.ci95.0\endcsname{$+$8.3}
\expandafter\def\csname val@rev.desc.d1.sspro.p2.C.zoomclick-d3.tau-target.ci95.1\endcsname{0.271}
\expandafter\def\csname cnt@rev.desc.d1.sspro.p2.C.zoomclick-d3.tau-target.ci95.1\endcsname{0}
\expandafter\def\csname pct@rev.desc.d1.sspro.p2.C.zoomclick-d3.tau-target.ci95.1\endcsname{27.1}
\expandafter\def\csname ppt@rev.desc.d1.sspro.p2.C.zoomclick-d3.tau-target.ci95.1\endcsname{$+$27.1}
\expandafter\def\csname val@rev.desc.d1.sspro.p2.C.zoomclick-d3.tau-target.mean\endcsname{0.167}
\expandafter\def\csname cnt@rev.desc.d1.sspro.p2.C.zoomclick-d3.tau-target.mean\endcsname{0}
\expandafter\def\csname pct@rev.desc.d1.sspro.p2.C.zoomclick-d3.tau-target.mean\endcsname{16.7}
\expandafter\def\csname ppt@rev.desc.d1.sspro.p2.C.zoomclick-d3.tau-target.mean\endcsname{$+$16.7}
\expandafter\def\csname val@rev.desc.d1.sspro.p2.C.zoomclick-d3.tau-target.n-apps\endcsname{6}
\expandafter\def\csname cnt@rev.desc.d1.sspro.p2.C.zoomclick-d3.tau-target.n-apps\endcsname{6}
\expandafter\def\csname pct@rev.desc.d1.sspro.p2.C.zoomclick-d3.tau-target.n-apps\endcsname{600.0}
\expandafter\def\csname ppt@rev.desc.d1.sspro.p2.C.zoomclick-d3.tau-target.n-apps\endcsname{$+$600.0}
\expandafter\def\csname val@rev.desc.d1.sspro.p2.C.zoomclick-d3.tau-target.tost1.ci90-inside\endcsname{0}
\expandafter\def\csname cnt@rev.desc.d1.sspro.p2.C.zoomclick-d3.tau-target.tost1.ci90-inside\endcsname{0}
\expandafter\def\csname pct@rev.desc.d1.sspro.p2.C.zoomclick-d3.tau-target.tost1.ci90-inside\endcsname{0.0}
\expandafter\def\csname ppt@rev.desc.d1.sspro.p2.C.zoomclick-d3.tau-target.tost1.ci90-inside\endcsname{0.0}
\expandafter\def\csname val@rev.desc.d1.sspro.p2.C.zoomclick-d3.tau-target.tost1.equivalent\endcsname{0}
\expandafter\def\csname cnt@rev.desc.d1.sspro.p2.C.zoomclick-d3.tau-target.tost1.equivalent\endcsname{0}
\expandafter\def\csname pct@rev.desc.d1.sspro.p2.C.zoomclick-d3.tau-target.tost1.equivalent\endcsname{0.0}
\expandafter\def\csname ppt@rev.desc.d1.sspro.p2.C.zoomclick-d3.tau-target.tost1.equivalent\endcsname{0.0}
\expandafter\def\csname val@rev.desc.d1.sspro.p2.C.zoomclick-d3.tau-target.tost1.p\endcsname{1.000}
\expandafter\def\csname cnt@rev.desc.d1.sspro.p2.C.zoomclick-d3.tau-target.tost1.p\endcsname{1}
\expandafter\def\csname pct@rev.desc.d1.sspro.p2.C.zoomclick-d3.tau-target.tost1.p\endcsname{100.0}
\expandafter\def\csname ppt@rev.desc.d1.sspro.p2.C.zoomclick-d3.tau-target.tost1.p\endcsname{$+$100.0}
\expandafter\def\csname val@rev.desc.d1.sspro.p2.C.zoomclick-d3.tau-target.tost5.ci90-inside\endcsname{0}
\expandafter\def\csname cnt@rev.desc.d1.sspro.p2.C.zoomclick-d3.tau-target.tost5.ci90-inside\endcsname{0}
\expandafter\def\csname pct@rev.desc.d1.sspro.p2.C.zoomclick-d3.tau-target.tost5.ci90-inside\endcsname{0.0}
\expandafter\def\csname ppt@rev.desc.d1.sspro.p2.C.zoomclick-d3.tau-target.tost5.ci90-inside\endcsname{0.0}
\expandafter\def\csname val@rev.desc.d1.sspro.p2.C.zoomclick-d3.tau-target.tost5.equivalent\endcsname{0}
\expandafter\def\csname cnt@rev.desc.d1.sspro.p2.C.zoomclick-d3.tau-target.tost5.equivalent\endcsname{0}
\expandafter\def\csname pct@rev.desc.d1.sspro.p2.C.zoomclick-d3.tau-target.tost5.equivalent\endcsname{0.0}
\expandafter\def\csname ppt@rev.desc.d1.sspro.p2.C.zoomclick-d3.tau-target.tost5.equivalent\endcsname{0.0}
\expandafter\def\csname val@rev.desc.d1.sspro.p2.C.zoomclick-d3.tau-target.tost5.p\endcsname{0.996}
\expandafter\def\csname cnt@rev.desc.d1.sspro.p2.C.zoomclick-d3.tau-target.tost5.p\endcsname{1}
\expandafter\def\csname pct@rev.desc.d1.sspro.p2.C.zoomclick-d3.tau-target.tost5.p\endcsname{99.6}
\expandafter\def\csname ppt@rev.desc.d1.sspro.p2.C.zoomclick-d3.tau-target.tost5.p\endcsname{$+$99.6}
\expandafter\def\csname val@rev.desc.d1.sspro.p2.C.zoomclick-d3.zero-estimate\endcsname{0}
\expandafter\def\csname cnt@rev.desc.d1.sspro.p2.C.zoomclick-d3.zero-estimate\endcsname{0}
\expandafter\def\csname pct@rev.desc.d1.sspro.p2.C.zoomclick-d3.zero-estimate\endcsname{0.0}
\expandafter\def\csname ppt@rev.desc.d1.sspro.p2.C.zoomclick-d3.zero-estimate\endcsname{0.0}
\expandafter\def\csname val@rev.desc.d1.sspro.p2.T.center-d1.delta-tau.ci90.0\endcsname{-0.010}
\expandafter\def\csname cnt@rev.desc.d1.sspro.p2.T.center-d1.delta-tau.ci90.0\endcsname{0}
\expandafter\def\csname pct@rev.desc.d1.sspro.p2.T.center-d1.delta-tau.ci90.0\endcsname{$-$1.0}
\expandafter\def\csname ppt@rev.desc.d1.sspro.p2.T.center-d1.delta-tau.ci90.0\endcsname{$-$1.0}
\expandafter\def\csname val@rev.desc.d1.sspro.p2.T.center-d1.delta-tau.ci90.1\endcsname{0.115}
\expandafter\def\csname cnt@rev.desc.d1.sspro.p2.T.center-d1.delta-tau.ci90.1\endcsname{0}
\expandafter\def\csname pct@rev.desc.d1.sspro.p2.T.center-d1.delta-tau.ci90.1\endcsname{11.5}
\expandafter\def\csname ppt@rev.desc.d1.sspro.p2.T.center-d1.delta-tau.ci90.1\endcsname{$+$11.5}
\expandafter\def\csname val@rev.desc.d1.sspro.p2.T.center-d1.delta-tau.ci95-excludes-0\endcsname{0}
\expandafter\def\csname cnt@rev.desc.d1.sspro.p2.T.center-d1.delta-tau.ci95-excludes-0\endcsname{0}
\expandafter\def\csname pct@rev.desc.d1.sspro.p2.T.center-d1.delta-tau.ci95-excludes-0\endcsname{0.0}
\expandafter\def\csname ppt@rev.desc.d1.sspro.p2.T.center-d1.delta-tau.ci95-excludes-0\endcsname{0.0}
\expandafter\def\csname val@rev.desc.d1.sspro.p2.T.center-d1.delta-tau.ci95.0\endcsname{-0.021}
\expandafter\def\csname cnt@rev.desc.d1.sspro.p2.T.center-d1.delta-tau.ci95.0\endcsname{0}
\expandafter\def\csname pct@rev.desc.d1.sspro.p2.T.center-d1.delta-tau.ci95.0\endcsname{$-$2.1}
\expandafter\def\csname ppt@rev.desc.d1.sspro.p2.T.center-d1.delta-tau.ci95.0\endcsname{$-$2.1}
\expandafter\def\csname val@rev.desc.d1.sspro.p2.T.center-d1.delta-tau.ci95.1\endcsname{0.125}
\expandafter\def\csname cnt@rev.desc.d1.sspro.p2.T.center-d1.delta-tau.ci95.1\endcsname{0}
\expandafter\def\csname pct@rev.desc.d1.sspro.p2.T.center-d1.delta-tau.ci95.1\endcsname{12.5}
\expandafter\def\csname ppt@rev.desc.d1.sspro.p2.T.center-d1.delta-tau.ci95.1\endcsname{$+$12.5}
\expandafter\def\csname val@rev.desc.d1.sspro.p2.T.center-d1.delta-tau.mean\endcsname{0.052}
\expandafter\def\csname cnt@rev.desc.d1.sspro.p2.T.center-d1.delta-tau.mean\endcsname{0}
\expandafter\def\csname pct@rev.desc.d1.sspro.p2.T.center-d1.delta-tau.mean\endcsname{5.2}
\expandafter\def\csname ppt@rev.desc.d1.sspro.p2.T.center-d1.delta-tau.mean\endcsname{$+$5.2}
\expandafter\def\csname val@rev.desc.d1.sspro.p2.T.center-d1.delta-tau.n-apps\endcsname{12}
\expandafter\def\csname cnt@rev.desc.d1.sspro.p2.T.center-d1.delta-tau.n-apps\endcsname{12}
\expandafter\def\csname pct@rev.desc.d1.sspro.p2.T.center-d1.delta-tau.n-apps\endcsname{1200.0}
\expandafter\def\csname ppt@rev.desc.d1.sspro.p2.T.center-d1.delta-tau.n-apps\endcsname{$+$1200.0}
\expandafter\def\csname val@rev.desc.d1.sspro.p2.T.center-d1.delta-tau.tost1.ci90-inside\endcsname{0}
\expandafter\def\csname cnt@rev.desc.d1.sspro.p2.T.center-d1.delta-tau.tost1.ci90-inside\endcsname{0}
\expandafter\def\csname pct@rev.desc.d1.sspro.p2.T.center-d1.delta-tau.tost1.ci90-inside\endcsname{0.0}
\expandafter\def\csname ppt@rev.desc.d1.sspro.p2.T.center-d1.delta-tau.tost1.ci90-inside\endcsname{0.0}
\expandafter\def\csname val@rev.desc.d1.sspro.p2.T.center-d1.delta-tau.tost1.equivalent\endcsname{0}
\expandafter\def\csname cnt@rev.desc.d1.sspro.p2.T.center-d1.delta-tau.tost1.equivalent\endcsname{0}
\expandafter\def\csname pct@rev.desc.d1.sspro.p2.T.center-d1.delta-tau.tost1.equivalent\endcsname{0.0}
\expandafter\def\csname ppt@rev.desc.d1.sspro.p2.T.center-d1.delta-tau.tost1.equivalent\endcsname{0.0}
\expandafter\def\csname val@rev.desc.d1.sspro.p2.T.center-d1.delta-tau.tost1.p\endcsname{0.899}
\expandafter\def\csname cnt@rev.desc.d1.sspro.p2.T.center-d1.delta-tau.tost1.p\endcsname{1}
\expandafter\def\csname pct@rev.desc.d1.sspro.p2.T.center-d1.delta-tau.tost1.p\endcsname{89.9}
\expandafter\def\csname ppt@rev.desc.d1.sspro.p2.T.center-d1.delta-tau.tost1.p\endcsname{$+$89.9}
\expandafter\def\csname val@rev.desc.d1.sspro.p2.T.center-d1.delta-tau.tost5.ci90-inside\endcsname{0}
\expandafter\def\csname cnt@rev.desc.d1.sspro.p2.T.center-d1.delta-tau.tost5.ci90-inside\endcsname{0}
\expandafter\def\csname pct@rev.desc.d1.sspro.p2.T.center-d1.delta-tau.tost5.ci90-inside\endcsname{0.0}
\expandafter\def\csname ppt@rev.desc.d1.sspro.p2.T.center-d1.delta-tau.tost5.ci90-inside\endcsname{0.0}
\expandafter\def\csname val@rev.desc.d1.sspro.p2.T.center-d1.delta-tau.tost5.equivalent\endcsname{0}
\expandafter\def\csname cnt@rev.desc.d1.sspro.p2.T.center-d1.delta-tau.tost5.equivalent\endcsname{0}
\expandafter\def\csname pct@rev.desc.d1.sspro.p2.T.center-d1.delta-tau.tost5.equivalent\endcsname{0.0}
\expandafter\def\csname ppt@rev.desc.d1.sspro.p2.T.center-d1.delta-tau.tost5.equivalent\endcsname{0.0}
\expandafter\def\csname val@rev.desc.d1.sspro.p2.T.center-d1.delta-tau.tost5.p\endcsname{0.553}
\expandafter\def\csname cnt@rev.desc.d1.sspro.p2.T.center-d1.delta-tau.tost5.p\endcsname{1}
\expandafter\def\csname pct@rev.desc.d1.sspro.p2.T.center-d1.delta-tau.tost5.p\endcsname{55.3}
\expandafter\def\csname ppt@rev.desc.d1.sspro.p2.T.center-d1.delta-tau.tost5.p\endcsname{$+$55.3}
\expandafter\def\csname val@rev.desc.d1.sspro.p2.T.center-d1.material-flip\endcsname{0}
\expandafter\def\csname cnt@rev.desc.d1.sspro.p2.T.center-d1.material-flip\endcsname{0}
\expandafter\def\csname pct@rev.desc.d1.sspro.p2.T.center-d1.material-flip\endcsname{0.0}
\expandafter\def\csname ppt@rev.desc.d1.sspro.p2.T.center-d1.material-flip\endcsname{0.0}
\expandafter\def\csname val@rev.desc.d1.sspro.p2.T.center-d1.n-apps\endcsname{12}
\expandafter\def\csname cnt@rev.desc.d1.sspro.p2.T.center-d1.n-apps\endcsname{12}
\expandafter\def\csname pct@rev.desc.d1.sspro.p2.T.center-d1.n-apps\endcsname{1200.0}
\expandafter\def\csname ppt@rev.desc.d1.sspro.p2.T.center-d1.n-apps\endcsname{$+$1200.0}
\expandafter\def\csname val@rev.desc.d1.sspro.p2.T.center-d1.n-tasks\endcsname{96}
\expandafter\def\csname cnt@rev.desc.d1.sspro.p2.T.center-d1.n-tasks\endcsname{96}
\expandafter\def\csname pct@rev.desc.d1.sspro.p2.T.center-d1.n-tasks\endcsname{9600.0}
\expandafter\def\csname ppt@rev.desc.d1.sspro.p2.T.center-d1.n-tasks\endcsname{$+$9600.0}
\expandafter\def\csname val@rev.desc.d1.sspro.p2.T.center-d1.sign-differs\endcsname{0}
\expandafter\def\csname cnt@rev.desc.d1.sspro.p2.T.center-d1.sign-differs\endcsname{0}
\expandafter\def\csname pct@rev.desc.d1.sspro.p2.T.center-d1.sign-differs\endcsname{0.0}
\expandafter\def\csname ppt@rev.desc.d1.sspro.p2.T.center-d1.sign-differs\endcsname{0.0}
\expandafter\def\csname val@rev.desc.d1.sspro.p2.T.center-d1.tau-source.ci90.0\endcsname{0.021}
\expandafter\def\csname cnt@rev.desc.d1.sspro.p2.T.center-d1.tau-source.ci90.0\endcsname{0}
\expandafter\def\csname pct@rev.desc.d1.sspro.p2.T.center-d1.tau-source.ci90.0\endcsname{2.1}
\expandafter\def\csname ppt@rev.desc.d1.sspro.p2.T.center-d1.tau-source.ci90.0\endcsname{$+$2.1}
\expandafter\def\csname val@rev.desc.d1.sspro.p2.T.center-d1.tau-source.ci90.1\endcsname{0.198}
\expandafter\def\csname cnt@rev.desc.d1.sspro.p2.T.center-d1.tau-source.ci90.1\endcsname{0}
\expandafter\def\csname pct@rev.desc.d1.sspro.p2.T.center-d1.tau-source.ci90.1\endcsname{19.8}
\expandafter\def\csname ppt@rev.desc.d1.sspro.p2.T.center-d1.tau-source.ci90.1\endcsname{$+$19.8}
\expandafter\def\csname val@rev.desc.d1.sspro.p2.T.center-d1.tau-source.ci95-excludes-0\endcsname{1}
\expandafter\def\csname cnt@rev.desc.d1.sspro.p2.T.center-d1.tau-source.ci95-excludes-0\endcsname{1}
\expandafter\def\csname pct@rev.desc.d1.sspro.p2.T.center-d1.tau-source.ci95-excludes-0\endcsname{100.0}
\expandafter\def\csname ppt@rev.desc.d1.sspro.p2.T.center-d1.tau-source.ci95-excludes-0\endcsname{$+$100.0}
\expandafter\def\csname val@rev.desc.d1.sspro.p2.T.center-d1.tau-source.ci95.0\endcsname{0.021}
\expandafter\def\csname cnt@rev.desc.d1.sspro.p2.T.center-d1.tau-source.ci95.0\endcsname{0}
\expandafter\def\csname pct@rev.desc.d1.sspro.p2.T.center-d1.tau-source.ci95.0\endcsname{2.1}
\expandafter\def\csname ppt@rev.desc.d1.sspro.p2.T.center-d1.tau-source.ci95.0\endcsname{$+$2.1}
\expandafter\def\csname val@rev.desc.d1.sspro.p2.T.center-d1.tau-source.ci95.1\endcsname{0.219}
\expandafter\def\csname cnt@rev.desc.d1.sspro.p2.T.center-d1.tau-source.ci95.1\endcsname{0}
\expandafter\def\csname pct@rev.desc.d1.sspro.p2.T.center-d1.tau-source.ci95.1\endcsname{21.9}
\expandafter\def\csname ppt@rev.desc.d1.sspro.p2.T.center-d1.tau-source.ci95.1\endcsname{$+$21.9}
\expandafter\def\csname val@rev.desc.d1.sspro.p2.T.center-d1.tau-source.mean\endcsname{0.104}
\expandafter\def\csname cnt@rev.desc.d1.sspro.p2.T.center-d1.tau-source.mean\endcsname{0}
\expandafter\def\csname pct@rev.desc.d1.sspro.p2.T.center-d1.tau-source.mean\endcsname{10.4}
\expandafter\def\csname ppt@rev.desc.d1.sspro.p2.T.center-d1.tau-source.mean\endcsname{$+$10.4}
\expandafter\def\csname val@rev.desc.d1.sspro.p2.T.center-d1.tau-source.n-apps\endcsname{12}
\expandafter\def\csname cnt@rev.desc.d1.sspro.p2.T.center-d1.tau-source.n-apps\endcsname{12}
\expandafter\def\csname pct@rev.desc.d1.sspro.p2.T.center-d1.tau-source.n-apps\endcsname{1200.0}
\expandafter\def\csname ppt@rev.desc.d1.sspro.p2.T.center-d1.tau-source.n-apps\endcsname{$+$1200.0}
\expandafter\def\csname val@rev.desc.d1.sspro.p2.T.center-d1.tau-source.tost1.ci90-inside\endcsname{0}
\expandafter\def\csname cnt@rev.desc.d1.sspro.p2.T.center-d1.tau-source.tost1.ci90-inside\endcsname{0}
\expandafter\def\csname pct@rev.desc.d1.sspro.p2.T.center-d1.tau-source.tost1.ci90-inside\endcsname{0.0}
\expandafter\def\csname ppt@rev.desc.d1.sspro.p2.T.center-d1.tau-source.tost1.ci90-inside\endcsname{0.0}
\expandafter\def\csname val@rev.desc.d1.sspro.p2.T.center-d1.tau-source.tost1.equivalent\endcsname{0}
\expandafter\def\csname cnt@rev.desc.d1.sspro.p2.T.center-d1.tau-source.tost1.equivalent\endcsname{0}
\expandafter\def\csname pct@rev.desc.d1.sspro.p2.T.center-d1.tau-source.tost1.equivalent\endcsname{0.0}
\expandafter\def\csname ppt@rev.desc.d1.sspro.p2.T.center-d1.tau-source.tost1.equivalent\endcsname{0.0}
\expandafter\def\csname val@rev.desc.d1.sspro.p2.T.center-d1.tau-source.tost1.p\endcsname{0.990}
\expandafter\def\csname cnt@rev.desc.d1.sspro.p2.T.center-d1.tau-source.tost1.p\endcsname{1}
\expandafter\def\csname pct@rev.desc.d1.sspro.p2.T.center-d1.tau-source.tost1.p\endcsname{99.0}
\expandafter\def\csname ppt@rev.desc.d1.sspro.p2.T.center-d1.tau-source.tost1.p\endcsname{$+$99.0}
\expandafter\def\csname val@rev.desc.d1.sspro.p2.T.center-d1.tau-source.tost5.ci90-inside\endcsname{0}
\expandafter\def\csname cnt@rev.desc.d1.sspro.p2.T.center-d1.tau-source.tost5.ci90-inside\endcsname{0}
\expandafter\def\csname pct@rev.desc.d1.sspro.p2.T.center-d1.tau-source.tost5.ci90-inside\endcsname{0.0}
\expandafter\def\csname ppt@rev.desc.d1.sspro.p2.T.center-d1.tau-source.tost5.ci90-inside\endcsname{0.0}
\expandafter\def\csname val@rev.desc.d1.sspro.p2.T.center-d1.tau-source.tost5.equivalent\endcsname{0}
\expandafter\def\csname cnt@rev.desc.d1.sspro.p2.T.center-d1.tau-source.tost5.equivalent\endcsname{0}
\expandafter\def\csname pct@rev.desc.d1.sspro.p2.T.center-d1.tau-source.tost5.equivalent\endcsname{0.0}
\expandafter\def\csname ppt@rev.desc.d1.sspro.p2.T.center-d1.tau-source.tost5.equivalent\endcsname{0.0}
\expandafter\def\csname val@rev.desc.d1.sspro.p2.T.center-d1.tau-source.tost5.p\endcsname{0.870}
\expandafter\def\csname cnt@rev.desc.d1.sspro.p2.T.center-d1.tau-source.tost5.p\endcsname{1}
\expandafter\def\csname pct@rev.desc.d1.sspro.p2.T.center-d1.tau-source.tost5.p\endcsname{87.0}
\expandafter\def\csname ppt@rev.desc.d1.sspro.p2.T.center-d1.tau-source.tost5.p\endcsname{$+$87.0}
\expandafter\def\csname val@rev.desc.d1.sspro.p2.T.center-d1.tau-target.ci90.0\endcsname{0.094}
\expandafter\def\csname cnt@rev.desc.d1.sspro.p2.T.center-d1.tau-target.ci90.0\endcsname{0}
\expandafter\def\csname pct@rev.desc.d1.sspro.p2.T.center-d1.tau-target.ci90.0\endcsname{9.4}
\expandafter\def\csname ppt@rev.desc.d1.sspro.p2.T.center-d1.tau-target.ci90.0\endcsname{$+$9.4}
\expandafter\def\csname val@rev.desc.d1.sspro.p2.T.center-d1.tau-target.ci90.1\endcsname{0.229}
\expandafter\def\csname cnt@rev.desc.d1.sspro.p2.T.center-d1.tau-target.ci90.1\endcsname{0}
\expandafter\def\csname pct@rev.desc.d1.sspro.p2.T.center-d1.tau-target.ci90.1\endcsname{22.9}
\expandafter\def\csname ppt@rev.desc.d1.sspro.p2.T.center-d1.tau-target.ci90.1\endcsname{$+$22.9}
\expandafter\def\csname val@rev.desc.d1.sspro.p2.T.center-d1.tau-target.ci95-excludes-0\endcsname{1}
\expandafter\def\csname cnt@rev.desc.d1.sspro.p2.T.center-d1.tau-target.ci95-excludes-0\endcsname{1}
\expandafter\def\csname pct@rev.desc.d1.sspro.p2.T.center-d1.tau-target.ci95-excludes-0\endcsname{100.0}
\expandafter\def\csname ppt@rev.desc.d1.sspro.p2.T.center-d1.tau-target.ci95-excludes-0\endcsname{$+$100.0}
\expandafter\def\csname val@rev.desc.d1.sspro.p2.T.center-d1.tau-target.ci95.0\endcsname{0.083}
\expandafter\def\csname cnt@rev.desc.d1.sspro.p2.T.center-d1.tau-target.ci95.0\endcsname{0}
\expandafter\def\csname pct@rev.desc.d1.sspro.p2.T.center-d1.tau-target.ci95.0\endcsname{8.3}
\expandafter\def\csname ppt@rev.desc.d1.sspro.p2.T.center-d1.tau-target.ci95.0\endcsname{$+$8.3}
\expandafter\def\csname val@rev.desc.d1.sspro.p2.T.center-d1.tau-target.ci95.1\endcsname{0.240}
\expandafter\def\csname cnt@rev.desc.d1.sspro.p2.T.center-d1.tau-target.ci95.1\endcsname{0}
\expandafter\def\csname pct@rev.desc.d1.sspro.p2.T.center-d1.tau-target.ci95.1\endcsname{24.0}
\expandafter\def\csname ppt@rev.desc.d1.sspro.p2.T.center-d1.tau-target.ci95.1\endcsname{$+$24.0}
\expandafter\def\csname val@rev.desc.d1.sspro.p2.T.center-d1.tau-target.mean\endcsname{0.156}
\expandafter\def\csname cnt@rev.desc.d1.sspro.p2.T.center-d1.tau-target.mean\endcsname{0}
\expandafter\def\csname pct@rev.desc.d1.sspro.p2.T.center-d1.tau-target.mean\endcsname{15.6}
\expandafter\def\csname ppt@rev.desc.d1.sspro.p2.T.center-d1.tau-target.mean\endcsname{$+$15.6}
\expandafter\def\csname val@rev.desc.d1.sspro.p2.T.center-d1.tau-target.n-apps\endcsname{12}
\expandafter\def\csname cnt@rev.desc.d1.sspro.p2.T.center-d1.tau-target.n-apps\endcsname{12}
\expandafter\def\csname pct@rev.desc.d1.sspro.p2.T.center-d1.tau-target.n-apps\endcsname{1200.0}
\expandafter\def\csname ppt@rev.desc.d1.sspro.p2.T.center-d1.tau-target.n-apps\endcsname{$+$1200.0}
\expandafter\def\csname val@rev.desc.d1.sspro.p2.T.center-d1.tau-target.tost1.ci90-inside\endcsname{0}
\expandafter\def\csname cnt@rev.desc.d1.sspro.p2.T.center-d1.tau-target.tost1.ci90-inside\endcsname{0}
\expandafter\def\csname pct@rev.desc.d1.sspro.p2.T.center-d1.tau-target.tost1.ci90-inside\endcsname{0.0}
\expandafter\def\csname ppt@rev.desc.d1.sspro.p2.T.center-d1.tau-target.tost1.ci90-inside\endcsname{0.0}
\expandafter\def\csname val@rev.desc.d1.sspro.p2.T.center-d1.tau-target.tost1.equivalent\endcsname{0}
\expandafter\def\csname cnt@rev.desc.d1.sspro.p2.T.center-d1.tau-target.tost1.equivalent\endcsname{0}
\expandafter\def\csname pct@rev.desc.d1.sspro.p2.T.center-d1.tau-target.tost1.equivalent\endcsname{0.0}
\expandafter\def\csname ppt@rev.desc.d1.sspro.p2.T.center-d1.tau-target.tost1.equivalent\endcsname{0.0}
\expandafter\def\csname val@rev.desc.d1.sspro.p2.T.center-d1.tau-target.tost1.p\endcsname{1.000}
\expandafter\def\csname cnt@rev.desc.d1.sspro.p2.T.center-d1.tau-target.tost1.p\endcsname{1}
\expandafter\def\csname pct@rev.desc.d1.sspro.p2.T.center-d1.tau-target.tost1.p\endcsname{100.0}
\expandafter\def\csname ppt@rev.desc.d1.sspro.p2.T.center-d1.tau-target.tost1.p\endcsname{$+$100.0}
\expandafter\def\csname val@rev.desc.d1.sspro.p2.T.center-d1.tau-target.tost5.ci90-inside\endcsname{0}
\expandafter\def\csname cnt@rev.desc.d1.sspro.p2.T.center-d1.tau-target.tost5.ci90-inside\endcsname{0}
\expandafter\def\csname pct@rev.desc.d1.sspro.p2.T.center-d1.tau-target.tost5.ci90-inside\endcsname{0.0}
\expandafter\def\csname ppt@rev.desc.d1.sspro.p2.T.center-d1.tau-target.tost5.ci90-inside\endcsname{0.0}
\expandafter\def\csname val@rev.desc.d1.sspro.p2.T.center-d1.tau-target.tost5.equivalent\endcsname{0}
\expandafter\def\csname cnt@rev.desc.d1.sspro.p2.T.center-d1.tau-target.tost5.equivalent\endcsname{0}
\expandafter\def\csname pct@rev.desc.d1.sspro.p2.T.center-d1.tau-target.tost5.equivalent\endcsname{0.0}
\expandafter\def\csname ppt@rev.desc.d1.sspro.p2.T.center-d1.tau-target.tost5.equivalent\endcsname{0.0}
\expandafter\def\csname val@rev.desc.d1.sspro.p2.T.center-d1.tau-target.tost5.p\endcsname{0.999}
\expandafter\def\csname cnt@rev.desc.d1.sspro.p2.T.center-d1.tau-target.tost5.p\endcsname{1}
\expandafter\def\csname pct@rev.desc.d1.sspro.p2.T.center-d1.tau-target.tost5.p\endcsname{99.9}
\expandafter\def\csname ppt@rev.desc.d1.sspro.p2.T.center-d1.tau-target.tost5.p\endcsname{$+$99.9}
\expandafter\def\csname val@rev.desc.d1.sspro.p2.T.center-d1.zero-estimate\endcsname{0}
\expandafter\def\csname cnt@rev.desc.d1.sspro.p2.T.center-d1.zero-estimate\endcsname{0}
\expandafter\def\csname pct@rev.desc.d1.sspro.p2.T.center-d1.zero-estimate\endcsname{0.0}
\expandafter\def\csname ppt@rev.desc.d1.sspro.p2.T.center-d1.zero-estimate\endcsname{0.0}
\expandafter\def\csname val@rev.desc.d1.sspro.p2.T.context-d1.delta-tau.ci90.0\endcsname{-0.052}
\expandafter\def\csname cnt@rev.desc.d1.sspro.p2.T.context-d1.delta-tau.ci90.0\endcsname{0}
\expandafter\def\csname pct@rev.desc.d1.sspro.p2.T.context-d1.delta-tau.ci90.0\endcsname{$-$5.2}
\expandafter\def\csname ppt@rev.desc.d1.sspro.p2.T.context-d1.delta-tau.ci90.0\endcsname{$-$5.2}
\expandafter\def\csname val@rev.desc.d1.sspro.p2.T.context-d1.delta-tau.ci90.1\endcsname{0.073}
\expandafter\def\csname cnt@rev.desc.d1.sspro.p2.T.context-d1.delta-tau.ci90.1\endcsname{0}
\expandafter\def\csname pct@rev.desc.d1.sspro.p2.T.context-d1.delta-tau.ci90.1\endcsname{7.3}
\expandafter\def\csname ppt@rev.desc.d1.sspro.p2.T.context-d1.delta-tau.ci90.1\endcsname{$+$7.3}
\expandafter\def\csname val@rev.desc.d1.sspro.p2.T.context-d1.delta-tau.ci95-excludes-0\endcsname{0}
\expandafter\def\csname cnt@rev.desc.d1.sspro.p2.T.context-d1.delta-tau.ci95-excludes-0\endcsname{0}
\expandafter\def\csname pct@rev.desc.d1.sspro.p2.T.context-d1.delta-tau.ci95-excludes-0\endcsname{0.0}
\expandafter\def\csname ppt@rev.desc.d1.sspro.p2.T.context-d1.delta-tau.ci95-excludes-0\endcsname{0.0}
\expandafter\def\csname val@rev.desc.d1.sspro.p2.T.context-d1.delta-tau.ci95.0\endcsname{-0.062}
\expandafter\def\csname cnt@rev.desc.d1.sspro.p2.T.context-d1.delta-tau.ci95.0\endcsname{0}
\expandafter\def\csname pct@rev.desc.d1.sspro.p2.T.context-d1.delta-tau.ci95.0\endcsname{$-$6.2}
\expandafter\def\csname ppt@rev.desc.d1.sspro.p2.T.context-d1.delta-tau.ci95.0\endcsname{$-$6.2}
\expandafter\def\csname val@rev.desc.d1.sspro.p2.T.context-d1.delta-tau.ci95.1\endcsname{0.083}
\expandafter\def\csname cnt@rev.desc.d1.sspro.p2.T.context-d1.delta-tau.ci95.1\endcsname{0}
\expandafter\def\csname pct@rev.desc.d1.sspro.p2.T.context-d1.delta-tau.ci95.1\endcsname{8.3}
\expandafter\def\csname ppt@rev.desc.d1.sspro.p2.T.context-d1.delta-tau.ci95.1\endcsname{$+$8.3}
\expandafter\def\csname val@rev.desc.d1.sspro.p2.T.context-d1.delta-tau.mean\endcsname{0.010}
\expandafter\def\csname cnt@rev.desc.d1.sspro.p2.T.context-d1.delta-tau.mean\endcsname{0}
\expandafter\def\csname pct@rev.desc.d1.sspro.p2.T.context-d1.delta-tau.mean\endcsname{1.0}
\expandafter\def\csname ppt@rev.desc.d1.sspro.p2.T.context-d1.delta-tau.mean\endcsname{$+$1.0}
\expandafter\def\csname val@rev.desc.d1.sspro.p2.T.context-d1.delta-tau.n-apps\endcsname{12}
\expandafter\def\csname cnt@rev.desc.d1.sspro.p2.T.context-d1.delta-tau.n-apps\endcsname{12}
\expandafter\def\csname pct@rev.desc.d1.sspro.p2.T.context-d1.delta-tau.n-apps\endcsname{1200.0}
\expandafter\def\csname ppt@rev.desc.d1.sspro.p2.T.context-d1.delta-tau.n-apps\endcsname{$+$1200.0}
\expandafter\def\csname val@rev.desc.d1.sspro.p2.T.context-d1.delta-tau.tost1.ci90-inside\endcsname{0}
\expandafter\def\csname cnt@rev.desc.d1.sspro.p2.T.context-d1.delta-tau.tost1.ci90-inside\endcsname{0}
\expandafter\def\csname pct@rev.desc.d1.sspro.p2.T.context-d1.delta-tau.tost1.ci90-inside\endcsname{0.0}
\expandafter\def\csname ppt@rev.desc.d1.sspro.p2.T.context-d1.delta-tau.tost1.ci90-inside\endcsname{0.0}
\expandafter\def\csname val@rev.desc.d1.sspro.p2.T.context-d1.delta-tau.tost1.equivalent\endcsname{0}
\expandafter\def\csname cnt@rev.desc.d1.sspro.p2.T.context-d1.delta-tau.tost1.equivalent\endcsname{0}
\expandafter\def\csname pct@rev.desc.d1.sspro.p2.T.context-d1.delta-tau.tost1.equivalent\endcsname{0.0}
\expandafter\def\csname ppt@rev.desc.d1.sspro.p2.T.context-d1.delta-tau.tost1.equivalent\endcsname{0.0}
\expandafter\def\csname val@rev.desc.d1.sspro.p2.T.context-d1.delta-tau.tost1.p\endcsname{0.563}
\expandafter\def\csname cnt@rev.desc.d1.sspro.p2.T.context-d1.delta-tau.tost1.p\endcsname{1}
\expandafter\def\csname pct@rev.desc.d1.sspro.p2.T.context-d1.delta-tau.tost1.p\endcsname{56.3}
\expandafter\def\csname ppt@rev.desc.d1.sspro.p2.T.context-d1.delta-tau.tost1.p\endcsname{$+$56.3}
\expandafter\def\csname val@rev.desc.d1.sspro.p2.T.context-d1.delta-tau.tost5.ci90-inside\endcsname{0}
\expandafter\def\csname cnt@rev.desc.d1.sspro.p2.T.context-d1.delta-tau.tost5.ci90-inside\endcsname{0}
\expandafter\def\csname pct@rev.desc.d1.sspro.p2.T.context-d1.delta-tau.tost5.ci90-inside\endcsname{0.0}
\expandafter\def\csname ppt@rev.desc.d1.sspro.p2.T.context-d1.delta-tau.tost5.ci90-inside\endcsname{0.0}
\expandafter\def\csname val@rev.desc.d1.sspro.p2.T.context-d1.delta-tau.tost5.equivalent\endcsname{0}
\expandafter\def\csname cnt@rev.desc.d1.sspro.p2.T.context-d1.delta-tau.tost5.equivalent\endcsname{0}
\expandafter\def\csname pct@rev.desc.d1.sspro.p2.T.context-d1.delta-tau.tost5.equivalent\endcsname{0.0}
\expandafter\def\csname ppt@rev.desc.d1.sspro.p2.T.context-d1.delta-tau.tost5.equivalent\endcsname{0.0}
\expandafter\def\csname val@rev.desc.d1.sspro.p2.T.context-d1.delta-tau.tost5.p\endcsname{0.170}
\expandafter\def\csname cnt@rev.desc.d1.sspro.p2.T.context-d1.delta-tau.tost5.p\endcsname{0}
\expandafter\def\csname pct@rev.desc.d1.sspro.p2.T.context-d1.delta-tau.tost5.p\endcsname{17.0}
\expandafter\def\csname ppt@rev.desc.d1.sspro.p2.T.context-d1.delta-tau.tost5.p\endcsname{$+$17.0}
\expandafter\def\csname val@rev.desc.d1.sspro.p2.T.context-d1.material-flip\endcsname{0}
\expandafter\def\csname cnt@rev.desc.d1.sspro.p2.T.context-d1.material-flip\endcsname{0}
\expandafter\def\csname pct@rev.desc.d1.sspro.p2.T.context-d1.material-flip\endcsname{0.0}
\expandafter\def\csname ppt@rev.desc.d1.sspro.p2.T.context-d1.material-flip\endcsname{0.0}
\expandafter\def\csname val@rev.desc.d1.sspro.p2.T.context-d1.n-apps\endcsname{12}
\expandafter\def\csname cnt@rev.desc.d1.sspro.p2.T.context-d1.n-apps\endcsname{12}
\expandafter\def\csname pct@rev.desc.d1.sspro.p2.T.context-d1.n-apps\endcsname{1200.0}
\expandafter\def\csname ppt@rev.desc.d1.sspro.p2.T.context-d1.n-apps\endcsname{$+$1200.0}
\expandafter\def\csname val@rev.desc.d1.sspro.p2.T.context-d1.n-tasks\endcsname{96}
\expandafter\def\csname cnt@rev.desc.d1.sspro.p2.T.context-d1.n-tasks\endcsname{96}
\expandafter\def\csname pct@rev.desc.d1.sspro.p2.T.context-d1.n-tasks\endcsname{9600.0}
\expandafter\def\csname ppt@rev.desc.d1.sspro.p2.T.context-d1.n-tasks\endcsname{$+$9600.0}
\expandafter\def\csname val@rev.desc.d1.sspro.p2.T.context-d1.sign-differs\endcsname{0}
\expandafter\def\csname cnt@rev.desc.d1.sspro.p2.T.context-d1.sign-differs\endcsname{0}
\expandafter\def\csname pct@rev.desc.d1.sspro.p2.T.context-d1.sign-differs\endcsname{0.0}
\expandafter\def\csname ppt@rev.desc.d1.sspro.p2.T.context-d1.sign-differs\endcsname{0.0}
\expandafter\def\csname val@rev.desc.d1.sspro.p2.T.context-d1.tau-source.ci90.0\endcsname{0.031}
\expandafter\def\csname cnt@rev.desc.d1.sspro.p2.T.context-d1.tau-source.ci90.0\endcsname{0}
\expandafter\def\csname pct@rev.desc.d1.sspro.p2.T.context-d1.tau-source.ci90.0\endcsname{3.1}
\expandafter\def\csname ppt@rev.desc.d1.sspro.p2.T.context-d1.tau-source.ci90.0\endcsname{$+$3.1}
\expandafter\def\csname val@rev.desc.d1.sspro.p2.T.context-d1.tau-source.ci90.1\endcsname{0.229}
\expandafter\def\csname cnt@rev.desc.d1.sspro.p2.T.context-d1.tau-source.ci90.1\endcsname{0}
\expandafter\def\csname pct@rev.desc.d1.sspro.p2.T.context-d1.tau-source.ci90.1\endcsname{22.9}
\expandafter\def\csname ppt@rev.desc.d1.sspro.p2.T.context-d1.tau-source.ci90.1\endcsname{$+$22.9}
\expandafter\def\csname val@rev.desc.d1.sspro.p2.T.context-d1.tau-source.ci95-excludes-0\endcsname{1}
\expandafter\def\csname cnt@rev.desc.d1.sspro.p2.T.context-d1.tau-source.ci95-excludes-0\endcsname{1}
\expandafter\def\csname pct@rev.desc.d1.sspro.p2.T.context-d1.tau-source.ci95-excludes-0\endcsname{100.0}
\expandafter\def\csname ppt@rev.desc.d1.sspro.p2.T.context-d1.tau-source.ci95-excludes-0\endcsname{$+$100.0}
\expandafter\def\csname val@rev.desc.d1.sspro.p2.T.context-d1.tau-source.ci95.0\endcsname{0.021}
\expandafter\def\csname cnt@rev.desc.d1.sspro.p2.T.context-d1.tau-source.ci95.0\endcsname{0}
\expandafter\def\csname pct@rev.desc.d1.sspro.p2.T.context-d1.tau-source.ci95.0\endcsname{2.1}
\expandafter\def\csname ppt@rev.desc.d1.sspro.p2.T.context-d1.tau-source.ci95.0\endcsname{$+$2.1}
\expandafter\def\csname val@rev.desc.d1.sspro.p2.T.context-d1.tau-source.ci95.1\endcsname{0.260}
\expandafter\def\csname cnt@rev.desc.d1.sspro.p2.T.context-d1.tau-source.ci95.1\endcsname{0}
\expandafter\def\csname pct@rev.desc.d1.sspro.p2.T.context-d1.tau-source.ci95.1\endcsname{26.0}
\expandafter\def\csname ppt@rev.desc.d1.sspro.p2.T.context-d1.tau-source.ci95.1\endcsname{$+$26.0}
\expandafter\def\csname val@rev.desc.d1.sspro.p2.T.context-d1.tau-source.mean\endcsname{0.115}
\expandafter\def\csname cnt@rev.desc.d1.sspro.p2.T.context-d1.tau-source.mean\endcsname{0}
\expandafter\def\csname pct@rev.desc.d1.sspro.p2.T.context-d1.tau-source.mean\endcsname{11.5}
\expandafter\def\csname ppt@rev.desc.d1.sspro.p2.T.context-d1.tau-source.mean\endcsname{$+$11.5}
\expandafter\def\csname val@rev.desc.d1.sspro.p2.T.context-d1.tau-source.n-apps\endcsname{12}
\expandafter\def\csname cnt@rev.desc.d1.sspro.p2.T.context-d1.tau-source.n-apps\endcsname{12}
\expandafter\def\csname pct@rev.desc.d1.sspro.p2.T.context-d1.tau-source.n-apps\endcsname{1200.0}
\expandafter\def\csname ppt@rev.desc.d1.sspro.p2.T.context-d1.tau-source.n-apps\endcsname{$+$1200.0}
\expandafter\def\csname val@rev.desc.d1.sspro.p2.T.context-d1.tau-source.tost1.ci90-inside\endcsname{0}
\expandafter\def\csname cnt@rev.desc.d1.sspro.p2.T.context-d1.tau-source.tost1.ci90-inside\endcsname{0}
\expandafter\def\csname pct@rev.desc.d1.sspro.p2.T.context-d1.tau-source.tost1.ci90-inside\endcsname{0.0}
\expandafter\def\csname ppt@rev.desc.d1.sspro.p2.T.context-d1.tau-source.tost1.ci90-inside\endcsname{0.0}
\expandafter\def\csname val@rev.desc.d1.sspro.p2.T.context-d1.tau-source.tost1.equivalent\endcsname{0}
\expandafter\def\csname cnt@rev.desc.d1.sspro.p2.T.context-d1.tau-source.tost1.equivalent\endcsname{0}
\expandafter\def\csname pct@rev.desc.d1.sspro.p2.T.context-d1.tau-source.tost1.equivalent\endcsname{0.0}
\expandafter\def\csname ppt@rev.desc.d1.sspro.p2.T.context-d1.tau-source.tost1.equivalent\endcsname{0.0}
\expandafter\def\csname val@rev.desc.d1.sspro.p2.T.context-d1.tau-source.tost1.p\endcsname{0.989}
\expandafter\def\csname cnt@rev.desc.d1.sspro.p2.T.context-d1.tau-source.tost1.p\endcsname{1}
\expandafter\def\csname pct@rev.desc.d1.sspro.p2.T.context-d1.tau-source.tost1.p\endcsname{98.9}
\expandafter\def\csname ppt@rev.desc.d1.sspro.p2.T.context-d1.tau-source.tost1.p\endcsname{$+$98.9}
\expandafter\def\csname val@rev.desc.d1.sspro.p2.T.context-d1.tau-source.tost5.ci90-inside\endcsname{0}
\expandafter\def\csname cnt@rev.desc.d1.sspro.p2.T.context-d1.tau-source.tost5.ci90-inside\endcsname{0}
\expandafter\def\csname pct@rev.desc.d1.sspro.p2.T.context-d1.tau-source.tost5.ci90-inside\endcsname{0.0}
\expandafter\def\csname ppt@rev.desc.d1.sspro.p2.T.context-d1.tau-source.tost5.ci90-inside\endcsname{0.0}
\expandafter\def\csname val@rev.desc.d1.sspro.p2.T.context-d1.tau-source.tost5.equivalent\endcsname{0}
\expandafter\def\csname cnt@rev.desc.d1.sspro.p2.T.context-d1.tau-source.tost5.equivalent\endcsname{0}
\expandafter\def\csname pct@rev.desc.d1.sspro.p2.T.context-d1.tau-source.tost5.equivalent\endcsname{0.0}
\expandafter\def\csname ppt@rev.desc.d1.sspro.p2.T.context-d1.tau-source.tost5.equivalent\endcsname{0.0}
\expandafter\def\csname val@rev.desc.d1.sspro.p2.T.context-d1.tau-source.tost5.p\endcsname{0.865}
\expandafter\def\csname cnt@rev.desc.d1.sspro.p2.T.context-d1.tau-source.tost5.p\endcsname{1}
\expandafter\def\csname pct@rev.desc.d1.sspro.p2.T.context-d1.tau-source.tost5.p\endcsname{86.5}
\expandafter\def\csname ppt@rev.desc.d1.sspro.p2.T.context-d1.tau-source.tost5.p\endcsname{$+$86.5}
\expandafter\def\csname val@rev.desc.d1.sspro.p2.T.context-d1.tau-target.ci90.0\endcsname{0.062}
\expandafter\def\csname cnt@rev.desc.d1.sspro.p2.T.context-d1.tau-target.ci90.0\endcsname{0}
\expandafter\def\csname pct@rev.desc.d1.sspro.p2.T.context-d1.tau-target.ci90.0\endcsname{6.2}
\expandafter\def\csname ppt@rev.desc.d1.sspro.p2.T.context-d1.tau-target.ci90.0\endcsname{$+$6.2}
\expandafter\def\csname val@rev.desc.d1.sspro.p2.T.context-d1.tau-target.ci90.1\endcsname{0.198}
\expandafter\def\csname cnt@rev.desc.d1.sspro.p2.T.context-d1.tau-target.ci90.1\endcsname{0}
\expandafter\def\csname pct@rev.desc.d1.sspro.p2.T.context-d1.tau-target.ci90.1\endcsname{19.8}
\expandafter\def\csname ppt@rev.desc.d1.sspro.p2.T.context-d1.tau-target.ci90.1\endcsname{$+$19.8}
\expandafter\def\csname val@rev.desc.d1.sspro.p2.T.context-d1.tau-target.ci95-excludes-0\endcsname{1}
\expandafter\def\csname cnt@rev.desc.d1.sspro.p2.T.context-d1.tau-target.ci95-excludes-0\endcsname{1}
\expandafter\def\csname pct@rev.desc.d1.sspro.p2.T.context-d1.tau-target.ci95-excludes-0\endcsname{100.0}
\expandafter\def\csname ppt@rev.desc.d1.sspro.p2.T.context-d1.tau-target.ci95-excludes-0\endcsname{$+$100.0}
\expandafter\def\csname val@rev.desc.d1.sspro.p2.T.context-d1.tau-target.ci95.0\endcsname{0.062}
\expandafter\def\csname cnt@rev.desc.d1.sspro.p2.T.context-d1.tau-target.ci95.0\endcsname{0}
\expandafter\def\csname pct@rev.desc.d1.sspro.p2.T.context-d1.tau-target.ci95.0\endcsname{6.2}
\expandafter\def\csname ppt@rev.desc.d1.sspro.p2.T.context-d1.tau-target.ci95.0\endcsname{$+$6.2}
\expandafter\def\csname val@rev.desc.d1.sspro.p2.T.context-d1.tau-target.ci95.1\endcsname{0.208}
\expandafter\def\csname cnt@rev.desc.d1.sspro.p2.T.context-d1.tau-target.ci95.1\endcsname{0}
\expandafter\def\csname pct@rev.desc.d1.sspro.p2.T.context-d1.tau-target.ci95.1\endcsname{20.8}
\expandafter\def\csname ppt@rev.desc.d1.sspro.p2.T.context-d1.tau-target.ci95.1\endcsname{$+$20.8}
\expandafter\def\csname val@rev.desc.d1.sspro.p2.T.context-d1.tau-target.mean\endcsname{0.125}
\expandafter\def\csname cnt@rev.desc.d1.sspro.p2.T.context-d1.tau-target.mean\endcsname{0}
\expandafter\def\csname pct@rev.desc.d1.sspro.p2.T.context-d1.tau-target.mean\endcsname{12.5}
\expandafter\def\csname ppt@rev.desc.d1.sspro.p2.T.context-d1.tau-target.mean\endcsname{$+$12.5}
\expandafter\def\csname val@rev.desc.d1.sspro.p2.T.context-d1.tau-target.n-apps\endcsname{12}
\expandafter\def\csname cnt@rev.desc.d1.sspro.p2.T.context-d1.tau-target.n-apps\endcsname{12}
\expandafter\def\csname pct@rev.desc.d1.sspro.p2.T.context-d1.tau-target.n-apps\endcsname{1200.0}
\expandafter\def\csname ppt@rev.desc.d1.sspro.p2.T.context-d1.tau-target.n-apps\endcsname{$+$1200.0}
\expandafter\def\csname val@rev.desc.d1.sspro.p2.T.context-d1.tau-target.tost1.ci90-inside\endcsname{0}
\expandafter\def\csname cnt@rev.desc.d1.sspro.p2.T.context-d1.tau-target.tost1.ci90-inside\endcsname{0}
\expandafter\def\csname pct@rev.desc.d1.sspro.p2.T.context-d1.tau-target.tost1.ci90-inside\endcsname{0.0}
\expandafter\def\csname ppt@rev.desc.d1.sspro.p2.T.context-d1.tau-target.tost1.ci90-inside\endcsname{0.0}
\expandafter\def\csname val@rev.desc.d1.sspro.p2.T.context-d1.tau-target.tost1.equivalent\endcsname{0}
\expandafter\def\csname cnt@rev.desc.d1.sspro.p2.T.context-d1.tau-target.tost1.equivalent\endcsname{0}
\expandafter\def\csname pct@rev.desc.d1.sspro.p2.T.context-d1.tau-target.tost1.equivalent\endcsname{0.0}
\expandafter\def\csname ppt@rev.desc.d1.sspro.p2.T.context-d1.tau-target.tost1.equivalent\endcsname{0.0}
\expandafter\def\csname val@rev.desc.d1.sspro.p2.T.context-d1.tau-target.tost1.p\endcsname{1.000}
\expandafter\def\csname cnt@rev.desc.d1.sspro.p2.T.context-d1.tau-target.tost1.p\endcsname{1}
\expandafter\def\csname pct@rev.desc.d1.sspro.p2.T.context-d1.tau-target.tost1.p\endcsname{100.0}
\expandafter\def\csname ppt@rev.desc.d1.sspro.p2.T.context-d1.tau-target.tost1.p\endcsname{$+$100.0}
\expandafter\def\csname val@rev.desc.d1.sspro.p2.T.context-d1.tau-target.tost5.ci90-inside\endcsname{0}
\expandafter\def\csname cnt@rev.desc.d1.sspro.p2.T.context-d1.tau-target.tost5.ci90-inside\endcsname{0}
\expandafter\def\csname pct@rev.desc.d1.sspro.p2.T.context-d1.tau-target.tost5.ci90-inside\endcsname{0.0}
\expandafter\def\csname ppt@rev.desc.d1.sspro.p2.T.context-d1.tau-target.tost5.ci90-inside\endcsname{0.0}
\expandafter\def\csname val@rev.desc.d1.sspro.p2.T.context-d1.tau-target.tost5.equivalent\endcsname{0}
\expandafter\def\csname cnt@rev.desc.d1.sspro.p2.T.context-d1.tau-target.tost5.equivalent\endcsname{0}
\expandafter\def\csname pct@rev.desc.d1.sspro.p2.T.context-d1.tau-target.tost5.equivalent\endcsname{0.0}
\expandafter\def\csname ppt@rev.desc.d1.sspro.p2.T.context-d1.tau-target.tost5.equivalent\endcsname{0.0}
\expandafter\def\csname val@rev.desc.d1.sspro.p2.T.context-d1.tau-target.tost5.p\endcsname{0.993}
\expandafter\def\csname cnt@rev.desc.d1.sspro.p2.T.context-d1.tau-target.tost5.p\endcsname{1}
\expandafter\def\csname pct@rev.desc.d1.sspro.p2.T.context-d1.tau-target.tost5.p\endcsname{99.3}
\expandafter\def\csname ppt@rev.desc.d1.sspro.p2.T.context-d1.tau-target.tost5.p\endcsname{$+$99.3}
\expandafter\def\csname val@rev.desc.d1.sspro.p2.T.context-d1.zero-estimate\endcsname{0}
\expandafter\def\csname cnt@rev.desc.d1.sspro.p2.T.context-d1.zero-estimate\endcsname{0}
\expandafter\def\csname pct@rev.desc.d1.sspro.p2.T.context-d1.zero-estimate\endcsname{0.0}
\expandafter\def\csname ppt@rev.desc.d1.sspro.p2.T.context-d1.zero-estimate\endcsname{0.0}
\expandafter\def\csname val@rev.desc.d1.sspro.p2.T.dual-d2.delta-tau.ci90.0\endcsname{-0.083}
\expandafter\def\csname cnt@rev.desc.d1.sspro.p2.T.dual-d2.delta-tau.ci90.0\endcsname{0}
\expandafter\def\csname pct@rev.desc.d1.sspro.p2.T.dual-d2.delta-tau.ci90.0\endcsname{$-$8.3}
\expandafter\def\csname ppt@rev.desc.d1.sspro.p2.T.dual-d2.delta-tau.ci90.0\endcsname{$-$8.3}
\expandafter\def\csname val@rev.desc.d1.sspro.p2.T.dual-d2.delta-tau.ci90.1\endcsname{0.094}
\expandafter\def\csname cnt@rev.desc.d1.sspro.p2.T.dual-d2.delta-tau.ci90.1\endcsname{0}
\expandafter\def\csname pct@rev.desc.d1.sspro.p2.T.dual-d2.delta-tau.ci90.1\endcsname{9.4}
\expandafter\def\csname ppt@rev.desc.d1.sspro.p2.T.dual-d2.delta-tau.ci90.1\endcsname{$+$9.4}
\expandafter\def\csname val@rev.desc.d1.sspro.p2.T.dual-d2.delta-tau.ci95-excludes-0\endcsname{0}
\expandafter\def\csname cnt@rev.desc.d1.sspro.p2.T.dual-d2.delta-tau.ci95-excludes-0\endcsname{0}
\expandafter\def\csname pct@rev.desc.d1.sspro.p2.T.dual-d2.delta-tau.ci95-excludes-0\endcsname{0.0}
\expandafter\def\csname ppt@rev.desc.d1.sspro.p2.T.dual-d2.delta-tau.ci95-excludes-0\endcsname{0.0}
\expandafter\def\csname val@rev.desc.d1.sspro.p2.T.dual-d2.delta-tau.ci95.0\endcsname{-0.104}
\expandafter\def\csname cnt@rev.desc.d1.sspro.p2.T.dual-d2.delta-tau.ci95.0\endcsname{0}
\expandafter\def\csname pct@rev.desc.d1.sspro.p2.T.dual-d2.delta-tau.ci95.0\endcsname{$-$10.4}
\expandafter\def\csname ppt@rev.desc.d1.sspro.p2.T.dual-d2.delta-tau.ci95.0\endcsname{$-$10.4}
\expandafter\def\csname val@rev.desc.d1.sspro.p2.T.dual-d2.delta-tau.ci95.1\endcsname{0.104}
\expandafter\def\csname cnt@rev.desc.d1.sspro.p2.T.dual-d2.delta-tau.ci95.1\endcsname{0}
\expandafter\def\csname pct@rev.desc.d1.sspro.p2.T.dual-d2.delta-tau.ci95.1\endcsname{10.4}
\expandafter\def\csname ppt@rev.desc.d1.sspro.p2.T.dual-d2.delta-tau.ci95.1\endcsname{$+$10.4}
\expandafter\def\csname val@rev.desc.d1.sspro.p2.T.dual-d2.delta-tau.mean\endcsname{0.010}
\expandafter\def\csname cnt@rev.desc.d1.sspro.p2.T.dual-d2.delta-tau.mean\endcsname{0}
\expandafter\def\csname pct@rev.desc.d1.sspro.p2.T.dual-d2.delta-tau.mean\endcsname{1.0}
\expandafter\def\csname ppt@rev.desc.d1.sspro.p2.T.dual-d2.delta-tau.mean\endcsname{$+$1.0}
\expandafter\def\csname val@rev.desc.d1.sspro.p2.T.dual-d2.delta-tau.n-apps\endcsname{12}
\expandafter\def\csname cnt@rev.desc.d1.sspro.p2.T.dual-d2.delta-tau.n-apps\endcsname{12}
\expandafter\def\csname pct@rev.desc.d1.sspro.p2.T.dual-d2.delta-tau.n-apps\endcsname{1200.0}
\expandafter\def\csname ppt@rev.desc.d1.sspro.p2.T.dual-d2.delta-tau.n-apps\endcsname{$+$1200.0}
\expandafter\def\csname val@rev.desc.d1.sspro.p2.T.dual-d2.delta-tau.tost1.ci90-inside\endcsname{0}
\expandafter\def\csname cnt@rev.desc.d1.sspro.p2.T.dual-d2.delta-tau.tost1.ci90-inside\endcsname{0}
\expandafter\def\csname pct@rev.desc.d1.sspro.p2.T.dual-d2.delta-tau.tost1.ci90-inside\endcsname{0.0}
\expandafter\def\csname ppt@rev.desc.d1.sspro.p2.T.dual-d2.delta-tau.tost1.ci90-inside\endcsname{0.0}
\expandafter\def\csname val@rev.desc.d1.sspro.p2.T.dual-d2.delta-tau.tost1.equivalent\endcsname{0}
\expandafter\def\csname cnt@rev.desc.d1.sspro.p2.T.dual-d2.delta-tau.tost1.equivalent\endcsname{0}
\expandafter\def\csname pct@rev.desc.d1.sspro.p2.T.dual-d2.delta-tau.tost1.equivalent\endcsname{0.0}
\expandafter\def\csname ppt@rev.desc.d1.sspro.p2.T.dual-d2.delta-tau.tost1.equivalent\endcsname{0.0}
\expandafter\def\csname val@rev.desc.d1.sspro.p2.T.dual-d2.delta-tau.tost1.p\endcsname{0.542}
\expandafter\def\csname cnt@rev.desc.d1.sspro.p2.T.dual-d2.delta-tau.tost1.p\endcsname{1}
\expandafter\def\csname pct@rev.desc.d1.sspro.p2.T.dual-d2.delta-tau.tost1.p\endcsname{54.2}
\expandafter\def\csname ppt@rev.desc.d1.sspro.p2.T.dual-d2.delta-tau.tost1.p\endcsname{$+$54.2}
\expandafter\def\csname val@rev.desc.d1.sspro.p2.T.dual-d2.delta-tau.tost5.ci90-inside\endcsname{0}
\expandafter\def\csname cnt@rev.desc.d1.sspro.p2.T.dual-d2.delta-tau.tost5.ci90-inside\endcsname{0}
\expandafter\def\csname pct@rev.desc.d1.sspro.p2.T.dual-d2.delta-tau.tost5.ci90-inside\endcsname{0.0}
\expandafter\def\csname ppt@rev.desc.d1.sspro.p2.T.dual-d2.delta-tau.tost5.ci90-inside\endcsname{0.0}
\expandafter\def\csname val@rev.desc.d1.sspro.p2.T.dual-d2.delta-tau.tost5.equivalent\endcsname{0}
\expandafter\def\csname cnt@rev.desc.d1.sspro.p2.T.dual-d2.delta-tau.tost5.equivalent\endcsname{0}
\expandafter\def\csname pct@rev.desc.d1.sspro.p2.T.dual-d2.delta-tau.tost5.equivalent\endcsname{0.0}
\expandafter\def\csname ppt@rev.desc.d1.sspro.p2.T.dual-d2.delta-tau.tost5.equivalent\endcsname{0.0}
\expandafter\def\csname val@rev.desc.d1.sspro.p2.T.dual-d2.delta-tau.tost5.p\endcsname{0.252}
\expandafter\def\csname cnt@rev.desc.d1.sspro.p2.T.dual-d2.delta-tau.tost5.p\endcsname{0}
\expandafter\def\csname pct@rev.desc.d1.sspro.p2.T.dual-d2.delta-tau.tost5.p\endcsname{25.2}
\expandafter\def\csname ppt@rev.desc.d1.sspro.p2.T.dual-d2.delta-tau.tost5.p\endcsname{$+$25.2}
\expandafter\def\csname val@rev.desc.d1.sspro.p2.T.dual-d2.material-flip\endcsname{0}
\expandafter\def\csname cnt@rev.desc.d1.sspro.p2.T.dual-d2.material-flip\endcsname{0}
\expandafter\def\csname pct@rev.desc.d1.sspro.p2.T.dual-d2.material-flip\endcsname{0.0}
\expandafter\def\csname ppt@rev.desc.d1.sspro.p2.T.dual-d2.material-flip\endcsname{0.0}
\expandafter\def\csname val@rev.desc.d1.sspro.p2.T.dual-d2.n-apps\endcsname{12}
\expandafter\def\csname cnt@rev.desc.d1.sspro.p2.T.dual-d2.n-apps\endcsname{12}
\expandafter\def\csname pct@rev.desc.d1.sspro.p2.T.dual-d2.n-apps\endcsname{1200.0}
\expandafter\def\csname ppt@rev.desc.d1.sspro.p2.T.dual-d2.n-apps\endcsname{$+$1200.0}
\expandafter\def\csname val@rev.desc.d1.sspro.p2.T.dual-d2.n-tasks\endcsname{96}
\expandafter\def\csname cnt@rev.desc.d1.sspro.p2.T.dual-d2.n-tasks\endcsname{96}
\expandafter\def\csname pct@rev.desc.d1.sspro.p2.T.dual-d2.n-tasks\endcsname{9600.0}
\expandafter\def\csname ppt@rev.desc.d1.sspro.p2.T.dual-d2.n-tasks\endcsname{$+$9600.0}
\expandafter\def\csname val@rev.desc.d1.sspro.p2.T.dual-d2.sign-differs\endcsname{0}
\expandafter\def\csname cnt@rev.desc.d1.sspro.p2.T.dual-d2.sign-differs\endcsname{0}
\expandafter\def\csname pct@rev.desc.d1.sspro.p2.T.dual-d2.sign-differs\endcsname{0.0}
\expandafter\def\csname ppt@rev.desc.d1.sspro.p2.T.dual-d2.sign-differs\endcsname{0.0}
\expandafter\def\csname val@rev.desc.d1.sspro.p2.T.dual-d2.tau-source.ci90.0\endcsname{0.031}
\expandafter\def\csname cnt@rev.desc.d1.sspro.p2.T.dual-d2.tau-source.ci90.0\endcsname{0}
\expandafter\def\csname pct@rev.desc.d1.sspro.p2.T.dual-d2.tau-source.ci90.0\endcsname{3.1}
\expandafter\def\csname ppt@rev.desc.d1.sspro.p2.T.dual-d2.tau-source.ci90.0\endcsname{$+$3.1}
\expandafter\def\csname val@rev.desc.d1.sspro.p2.T.dual-d2.tau-source.ci90.1\endcsname{0.271}
\expandafter\def\csname cnt@rev.desc.d1.sspro.p2.T.dual-d2.tau-source.ci90.1\endcsname{0}
\expandafter\def\csname pct@rev.desc.d1.sspro.p2.T.dual-d2.tau-source.ci90.1\endcsname{27.1}
\expandafter\def\csname ppt@rev.desc.d1.sspro.p2.T.dual-d2.tau-source.ci90.1\endcsname{$+$27.1}
\expandafter\def\csname val@rev.desc.d1.sspro.p2.T.dual-d2.tau-source.ci95-excludes-0\endcsname{1}
\expandafter\def\csname cnt@rev.desc.d1.sspro.p2.T.dual-d2.tau-source.ci95-excludes-0\endcsname{1}
\expandafter\def\csname pct@rev.desc.d1.sspro.p2.T.dual-d2.tau-source.ci95-excludes-0\endcsname{100.0}
\expandafter\def\csname ppt@rev.desc.d1.sspro.p2.T.dual-d2.tau-source.ci95-excludes-0\endcsname{$+$100.0}
\expandafter\def\csname val@rev.desc.d1.sspro.p2.T.dual-d2.tau-source.ci95.0\endcsname{0.010}
\expandafter\def\csname cnt@rev.desc.d1.sspro.p2.T.dual-d2.tau-source.ci95.0\endcsname{0}
\expandafter\def\csname pct@rev.desc.d1.sspro.p2.T.dual-d2.tau-source.ci95.0\endcsname{1.0}
\expandafter\def\csname ppt@rev.desc.d1.sspro.p2.T.dual-d2.tau-source.ci95.0\endcsname{$+$1.0}
\expandafter\def\csname val@rev.desc.d1.sspro.p2.T.dual-d2.tau-source.ci95.1\endcsname{0.302}
\expandafter\def\csname cnt@rev.desc.d1.sspro.p2.T.dual-d2.tau-source.ci95.1\endcsname{0}
\expandafter\def\csname pct@rev.desc.d1.sspro.p2.T.dual-d2.tau-source.ci95.1\endcsname{30.2}
\expandafter\def\csname ppt@rev.desc.d1.sspro.p2.T.dual-d2.tau-source.ci95.1\endcsname{$+$30.2}
\expandafter\def\csname val@rev.desc.d1.sspro.p2.T.dual-d2.tau-source.mean\endcsname{0.146}
\expandafter\def\csname cnt@rev.desc.d1.sspro.p2.T.dual-d2.tau-source.mean\endcsname{0}
\expandafter\def\csname pct@rev.desc.d1.sspro.p2.T.dual-d2.tau-source.mean\endcsname{14.6}
\expandafter\def\csname ppt@rev.desc.d1.sspro.p2.T.dual-d2.tau-source.mean\endcsname{$+$14.6}
\expandafter\def\csname val@rev.desc.d1.sspro.p2.T.dual-d2.tau-source.n-apps\endcsname{12}
\expandafter\def\csname cnt@rev.desc.d1.sspro.p2.T.dual-d2.tau-source.n-apps\endcsname{12}
\expandafter\def\csname pct@rev.desc.d1.sspro.p2.T.dual-d2.tau-source.n-apps\endcsname{1200.0}
\expandafter\def\csname ppt@rev.desc.d1.sspro.p2.T.dual-d2.tau-source.n-apps\endcsname{$+$1200.0}
\expandafter\def\csname val@rev.desc.d1.sspro.p2.T.dual-d2.tau-source.tost1.ci90-inside\endcsname{0}
\expandafter\def\csname cnt@rev.desc.d1.sspro.p2.T.dual-d2.tau-source.tost1.ci90-inside\endcsname{0}
\expandafter\def\csname pct@rev.desc.d1.sspro.p2.T.dual-d2.tau-source.tost1.ci90-inside\endcsname{0.0}
\expandafter\def\csname ppt@rev.desc.d1.sspro.p2.T.dual-d2.tau-source.tost1.ci90-inside\endcsname{0.0}
\expandafter\def\csname val@rev.desc.d1.sspro.p2.T.dual-d2.tau-source.tost1.equivalent\endcsname{0}
\expandafter\def\csname cnt@rev.desc.d1.sspro.p2.T.dual-d2.tau-source.tost1.equivalent\endcsname{0}
\expandafter\def\csname pct@rev.desc.d1.sspro.p2.T.dual-d2.tau-source.tost1.equivalent\endcsname{0.0}
\expandafter\def\csname ppt@rev.desc.d1.sspro.p2.T.dual-d2.tau-source.tost1.equivalent\endcsname{0.0}
\expandafter\def\csname val@rev.desc.d1.sspro.p2.T.dual-d2.tau-source.tost1.p\endcsname{0.980}
\expandafter\def\csname cnt@rev.desc.d1.sspro.p2.T.dual-d2.tau-source.tost1.p\endcsname{1}
\expandafter\def\csname pct@rev.desc.d1.sspro.p2.T.dual-d2.tau-source.tost1.p\endcsname{98.0}
\expandafter\def\csname ppt@rev.desc.d1.sspro.p2.T.dual-d2.tau-source.tost1.p\endcsname{$+$98.0}
\expandafter\def\csname val@rev.desc.d1.sspro.p2.T.dual-d2.tau-source.tost5.ci90-inside\endcsname{0}
\expandafter\def\csname cnt@rev.desc.d1.sspro.p2.T.dual-d2.tau-source.tost5.ci90-inside\endcsname{0}
\expandafter\def\csname pct@rev.desc.d1.sspro.p2.T.dual-d2.tau-source.tost5.ci90-inside\endcsname{0.0}
\expandafter\def\csname ppt@rev.desc.d1.sspro.p2.T.dual-d2.tau-source.tost5.ci90-inside\endcsname{0.0}
\expandafter\def\csname val@rev.desc.d1.sspro.p2.T.dual-d2.tau-source.tost5.equivalent\endcsname{0}
\expandafter\def\csname cnt@rev.desc.d1.sspro.p2.T.dual-d2.tau-source.tost5.equivalent\endcsname{0}
\expandafter\def\csname pct@rev.desc.d1.sspro.p2.T.dual-d2.tau-source.tost5.equivalent\endcsname{0.0}
\expandafter\def\csname ppt@rev.desc.d1.sspro.p2.T.dual-d2.tau-source.tost5.equivalent\endcsname{0.0}
\expandafter\def\csname val@rev.desc.d1.sspro.p2.T.dual-d2.tau-source.tost5.p\endcsname{0.920}
\expandafter\def\csname cnt@rev.desc.d1.sspro.p2.T.dual-d2.tau-source.tost5.p\endcsname{1}
\expandafter\def\csname pct@rev.desc.d1.sspro.p2.T.dual-d2.tau-source.tost5.p\endcsname{92.0}
\expandafter\def\csname ppt@rev.desc.d1.sspro.p2.T.dual-d2.tau-source.tost5.p\endcsname{$+$92.0}
\expandafter\def\csname val@rev.desc.d1.sspro.p2.T.dual-d2.tau-target.ci90.0\endcsname{0.094}
\expandafter\def\csname cnt@rev.desc.d1.sspro.p2.T.dual-d2.tau-target.ci90.0\endcsname{0}
\expandafter\def\csname pct@rev.desc.d1.sspro.p2.T.dual-d2.tau-target.ci90.0\endcsname{9.4}
\expandafter\def\csname ppt@rev.desc.d1.sspro.p2.T.dual-d2.tau-target.ci90.0\endcsname{$+$9.4}
\expandafter\def\csname val@rev.desc.d1.sspro.p2.T.dual-d2.tau-target.ci90.1\endcsname{0.219}
\expandafter\def\csname cnt@rev.desc.d1.sspro.p2.T.dual-d2.tau-target.ci90.1\endcsname{0}
\expandafter\def\csname pct@rev.desc.d1.sspro.p2.T.dual-d2.tau-target.ci90.1\endcsname{21.9}
\expandafter\def\csname ppt@rev.desc.d1.sspro.p2.T.dual-d2.tau-target.ci90.1\endcsname{$+$21.9}
\expandafter\def\csname val@rev.desc.d1.sspro.p2.T.dual-d2.tau-target.ci95-excludes-0\endcsname{1}
\expandafter\def\csname cnt@rev.desc.d1.sspro.p2.T.dual-d2.tau-target.ci95-excludes-0\endcsname{1}
\expandafter\def\csname pct@rev.desc.d1.sspro.p2.T.dual-d2.tau-target.ci95-excludes-0\endcsname{100.0}
\expandafter\def\csname ppt@rev.desc.d1.sspro.p2.T.dual-d2.tau-target.ci95-excludes-0\endcsname{$+$100.0}
\expandafter\def\csname val@rev.desc.d1.sspro.p2.T.dual-d2.tau-target.ci95.0\endcsname{0.083}
\expandafter\def\csname cnt@rev.desc.d1.sspro.p2.T.dual-d2.tau-target.ci95.0\endcsname{0}
\expandafter\def\csname pct@rev.desc.d1.sspro.p2.T.dual-d2.tau-target.ci95.0\endcsname{8.3}
\expandafter\def\csname ppt@rev.desc.d1.sspro.p2.T.dual-d2.tau-target.ci95.0\endcsname{$+$8.3}
\expandafter\def\csname val@rev.desc.d1.sspro.p2.T.dual-d2.tau-target.ci95.1\endcsname{0.229}
\expandafter\def\csname cnt@rev.desc.d1.sspro.p2.T.dual-d2.tau-target.ci95.1\endcsname{0}
\expandafter\def\csname pct@rev.desc.d1.sspro.p2.T.dual-d2.tau-target.ci95.1\endcsname{22.9}
\expandafter\def\csname ppt@rev.desc.d1.sspro.p2.T.dual-d2.tau-target.ci95.1\endcsname{$+$22.9}
\expandafter\def\csname val@rev.desc.d1.sspro.p2.T.dual-d2.tau-target.mean\endcsname{0.156}
\expandafter\def\csname cnt@rev.desc.d1.sspro.p2.T.dual-d2.tau-target.mean\endcsname{0}
\expandafter\def\csname pct@rev.desc.d1.sspro.p2.T.dual-d2.tau-target.mean\endcsname{15.6}
\expandafter\def\csname ppt@rev.desc.d1.sspro.p2.T.dual-d2.tau-target.mean\endcsname{$+$15.6}
\expandafter\def\csname val@rev.desc.d1.sspro.p2.T.dual-d2.tau-target.n-apps\endcsname{12}
\expandafter\def\csname cnt@rev.desc.d1.sspro.p2.T.dual-d2.tau-target.n-apps\endcsname{12}
\expandafter\def\csname pct@rev.desc.d1.sspro.p2.T.dual-d2.tau-target.n-apps\endcsname{1200.0}
\expandafter\def\csname ppt@rev.desc.d1.sspro.p2.T.dual-d2.tau-target.n-apps\endcsname{$+$1200.0}
\expandafter\def\csname val@rev.desc.d1.sspro.p2.T.dual-d2.tau-target.tost1.ci90-inside\endcsname{0}
\expandafter\def\csname cnt@rev.desc.d1.sspro.p2.T.dual-d2.tau-target.tost1.ci90-inside\endcsname{0}
\expandafter\def\csname pct@rev.desc.d1.sspro.p2.T.dual-d2.tau-target.tost1.ci90-inside\endcsname{0.0}
\expandafter\def\csname ppt@rev.desc.d1.sspro.p2.T.dual-d2.tau-target.tost1.ci90-inside\endcsname{0.0}
\expandafter\def\csname val@rev.desc.d1.sspro.p2.T.dual-d2.tau-target.tost1.equivalent\endcsname{0}
\expandafter\def\csname cnt@rev.desc.d1.sspro.p2.T.dual-d2.tau-target.tost1.equivalent\endcsname{0}
\expandafter\def\csname pct@rev.desc.d1.sspro.p2.T.dual-d2.tau-target.tost1.equivalent\endcsname{0.0}
\expandafter\def\csname ppt@rev.desc.d1.sspro.p2.T.dual-d2.tau-target.tost1.equivalent\endcsname{0.0}
\expandafter\def\csname val@rev.desc.d1.sspro.p2.T.dual-d2.tau-target.tost1.p\endcsname{1.000}
\expandafter\def\csname cnt@rev.desc.d1.sspro.p2.T.dual-d2.tau-target.tost1.p\endcsname{1}
\expandafter\def\csname pct@rev.desc.d1.sspro.p2.T.dual-d2.tau-target.tost1.p\endcsname{100.0}
\expandafter\def\csname ppt@rev.desc.d1.sspro.p2.T.dual-d2.tau-target.tost1.p\endcsname{$+$100.0}
\expandafter\def\csname val@rev.desc.d1.sspro.p2.T.dual-d2.tau-target.tost5.ci90-inside\endcsname{0}
\expandafter\def\csname cnt@rev.desc.d1.sspro.p2.T.dual-d2.tau-target.tost5.ci90-inside\endcsname{0}
\expandafter\def\csname pct@rev.desc.d1.sspro.p2.T.dual-d2.tau-target.tost5.ci90-inside\endcsname{0.0}
\expandafter\def\csname ppt@rev.desc.d1.sspro.p2.T.dual-d2.tau-target.tost5.ci90-inside\endcsname{0.0}
\expandafter\def\csname val@rev.desc.d1.sspro.p2.T.dual-d2.tau-target.tost5.equivalent\endcsname{0}
\expandafter\def\csname cnt@rev.desc.d1.sspro.p2.T.dual-d2.tau-target.tost5.equivalent\endcsname{0}
\expandafter\def\csname pct@rev.desc.d1.sspro.p2.T.dual-d2.tau-target.tost5.equivalent\endcsname{0.0}
\expandafter\def\csname ppt@rev.desc.d1.sspro.p2.T.dual-d2.tau-target.tost5.equivalent\endcsname{0.0}
\expandafter\def\csname val@rev.desc.d1.sspro.p2.T.dual-d2.tau-target.tost5.p\endcsname{1.000}
\expandafter\def\csname cnt@rev.desc.d1.sspro.p2.T.dual-d2.tau-target.tost5.p\endcsname{1}
\expandafter\def\csname pct@rev.desc.d1.sspro.p2.T.dual-d2.tau-target.tost5.p\endcsname{100.0}
\expandafter\def\csname ppt@rev.desc.d1.sspro.p2.T.dual-d2.tau-target.tost5.p\endcsname{$+$100.0}
\expandafter\def\csname val@rev.desc.d1.sspro.p2.T.dual-d2.zero-estimate\endcsname{0}
\expandafter\def\csname cnt@rev.desc.d1.sspro.p2.T.dual-d2.zero-estimate\endcsname{0}
\expandafter\def\csname pct@rev.desc.d1.sspro.p2.T.dual-d2.zero-estimate\endcsname{0.0}
\expandafter\def\csname ppt@rev.desc.d1.sspro.p2.T.dual-d2.zero-estimate\endcsname{0.0}
\expandafter\def\csname val@rev.desc.d1.sspro.p2.T.grid2-d3.delta-tau.ci90.0\endcsname{-0.042}
\expandafter\def\csname cnt@rev.desc.d1.sspro.p2.T.grid2-d3.delta-tau.ci90.0\endcsname{0}
\expandafter\def\csname pct@rev.desc.d1.sspro.p2.T.grid2-d3.delta-tau.ci90.0\endcsname{$-$4.2}
\expandafter\def\csname ppt@rev.desc.d1.sspro.p2.T.grid2-d3.delta-tau.ci90.0\endcsname{$-$4.2}
\expandafter\def\csname val@rev.desc.d1.sspro.p2.T.grid2-d3.delta-tau.ci90.1\endcsname{0.104}
\expandafter\def\csname cnt@rev.desc.d1.sspro.p2.T.grid2-d3.delta-tau.ci90.1\endcsname{0}
\expandafter\def\csname pct@rev.desc.d1.sspro.p2.T.grid2-d3.delta-tau.ci90.1\endcsname{10.4}
\expandafter\def\csname ppt@rev.desc.d1.sspro.p2.T.grid2-d3.delta-tau.ci90.1\endcsname{$+$10.4}
\expandafter\def\csname val@rev.desc.d1.sspro.p2.T.grid2-d3.delta-tau.ci95-excludes-0\endcsname{0}
\expandafter\def\csname cnt@rev.desc.d1.sspro.p2.T.grid2-d3.delta-tau.ci95-excludes-0\endcsname{0}
\expandafter\def\csname pct@rev.desc.d1.sspro.p2.T.grid2-d3.delta-tau.ci95-excludes-0\endcsname{0.0}
\expandafter\def\csname ppt@rev.desc.d1.sspro.p2.T.grid2-d3.delta-tau.ci95-excludes-0\endcsname{0.0}
\expandafter\def\csname val@rev.desc.d1.sspro.p2.T.grid2-d3.delta-tau.ci95.0\endcsname{-0.052}
\expandafter\def\csname cnt@rev.desc.d1.sspro.p2.T.grid2-d3.delta-tau.ci95.0\endcsname{0}
\expandafter\def\csname pct@rev.desc.d1.sspro.p2.T.grid2-d3.delta-tau.ci95.0\endcsname{$-$5.2}
\expandafter\def\csname ppt@rev.desc.d1.sspro.p2.T.grid2-d3.delta-tau.ci95.0\endcsname{$-$5.2}
\expandafter\def\csname val@rev.desc.d1.sspro.p2.T.grid2-d3.delta-tau.ci95.1\endcsname{0.115}
\expandafter\def\csname cnt@rev.desc.d1.sspro.p2.T.grid2-d3.delta-tau.ci95.1\endcsname{0}
\expandafter\def\csname pct@rev.desc.d1.sspro.p2.T.grid2-d3.delta-tau.ci95.1\endcsname{11.5}
\expandafter\def\csname ppt@rev.desc.d1.sspro.p2.T.grid2-d3.delta-tau.ci95.1\endcsname{$+$11.5}
\expandafter\def\csname val@rev.desc.d1.sspro.p2.T.grid2-d3.delta-tau.mean\endcsname{0.031}
\expandafter\def\csname cnt@rev.desc.d1.sspro.p2.T.grid2-d3.delta-tau.mean\endcsname{0}
\expandafter\def\csname pct@rev.desc.d1.sspro.p2.T.grid2-d3.delta-tau.mean\endcsname{3.1}
\expandafter\def\csname ppt@rev.desc.d1.sspro.p2.T.grid2-d3.delta-tau.mean\endcsname{$+$3.1}
\expandafter\def\csname val@rev.desc.d1.sspro.p2.T.grid2-d3.delta-tau.n-apps\endcsname{12}
\expandafter\def\csname cnt@rev.desc.d1.sspro.p2.T.grid2-d3.delta-tau.n-apps\endcsname{12}
\expandafter\def\csname pct@rev.desc.d1.sspro.p2.T.grid2-d3.delta-tau.n-apps\endcsname{1200.0}
\expandafter\def\csname ppt@rev.desc.d1.sspro.p2.T.grid2-d3.delta-tau.n-apps\endcsname{$+$1200.0}
\expandafter\def\csname val@rev.desc.d1.sspro.p2.T.grid2-d3.delta-tau.tost1.ci90-inside\endcsname{0}
\expandafter\def\csname cnt@rev.desc.d1.sspro.p2.T.grid2-d3.delta-tau.tost1.ci90-inside\endcsname{0}
\expandafter\def\csname pct@rev.desc.d1.sspro.p2.T.grid2-d3.delta-tau.tost1.ci90-inside\endcsname{0.0}
\expandafter\def\csname ppt@rev.desc.d1.sspro.p2.T.grid2-d3.delta-tau.tost1.ci90-inside\endcsname{0.0}
\expandafter\def\csname val@rev.desc.d1.sspro.p2.T.grid2-d3.delta-tau.tost1.equivalent\endcsname{0}
\expandafter\def\csname cnt@rev.desc.d1.sspro.p2.T.grid2-d3.delta-tau.tost1.equivalent\endcsname{0}
\expandafter\def\csname pct@rev.desc.d1.sspro.p2.T.grid2-d3.delta-tau.tost1.equivalent\endcsname{0.0}
\expandafter\def\csname ppt@rev.desc.d1.sspro.p2.T.grid2-d3.delta-tau.tost1.equivalent\endcsname{0.0}
\expandafter\def\csname val@rev.desc.d1.sspro.p2.T.grid2-d3.delta-tau.tost1.p\endcsname{0.732}
\expandafter\def\csname cnt@rev.desc.d1.sspro.p2.T.grid2-d3.delta-tau.tost1.p\endcsname{1}
\expandafter\def\csname pct@rev.desc.d1.sspro.p2.T.grid2-d3.delta-tau.tost1.p\endcsname{73.2}
\expandafter\def\csname ppt@rev.desc.d1.sspro.p2.T.grid2-d3.delta-tau.tost1.p\endcsname{$+$73.2}
\expandafter\def\csname val@rev.desc.d1.sspro.p2.T.grid2-d3.delta-tau.tost5.ci90-inside\endcsname{0}
\expandafter\def\csname cnt@rev.desc.d1.sspro.p2.T.grid2-d3.delta-tau.tost5.ci90-inside\endcsname{0}
\expandafter\def\csname pct@rev.desc.d1.sspro.p2.T.grid2-d3.delta-tau.tost5.ci90-inside\endcsname{0.0}
\expandafter\def\csname ppt@rev.desc.d1.sspro.p2.T.grid2-d3.delta-tau.tost5.ci90-inside\endcsname{0.0}
\expandafter\def\csname val@rev.desc.d1.sspro.p2.T.grid2-d3.delta-tau.tost5.equivalent\endcsname{0}
\expandafter\def\csname cnt@rev.desc.d1.sspro.p2.T.grid2-d3.delta-tau.tost5.equivalent\endcsname{0}
\expandafter\def\csname pct@rev.desc.d1.sspro.p2.T.grid2-d3.delta-tau.tost5.equivalent\endcsname{0.0}
\expandafter\def\csname ppt@rev.desc.d1.sspro.p2.T.grid2-d3.delta-tau.tost5.equivalent\endcsname{0.0}
\expandafter\def\csname val@rev.desc.d1.sspro.p2.T.grid2-d3.delta-tau.tost5.p\endcsname{0.357}
\expandafter\def\csname cnt@rev.desc.d1.sspro.p2.T.grid2-d3.delta-tau.tost5.p\endcsname{0}
\expandafter\def\csname pct@rev.desc.d1.sspro.p2.T.grid2-d3.delta-tau.tost5.p\endcsname{35.7}
\expandafter\def\csname ppt@rev.desc.d1.sspro.p2.T.grid2-d3.delta-tau.tost5.p\endcsname{$+$35.7}
\expandafter\def\csname val@rev.desc.d1.sspro.p2.T.grid2-d3.material-flip\endcsname{0}
\expandafter\def\csname cnt@rev.desc.d1.sspro.p2.T.grid2-d3.material-flip\endcsname{0}
\expandafter\def\csname pct@rev.desc.d1.sspro.p2.T.grid2-d3.material-flip\endcsname{0.0}
\expandafter\def\csname ppt@rev.desc.d1.sspro.p2.T.grid2-d3.material-flip\endcsname{0.0}
\expandafter\def\csname val@rev.desc.d1.sspro.p2.T.grid2-d3.n-apps\endcsname{12}
\expandafter\def\csname cnt@rev.desc.d1.sspro.p2.T.grid2-d3.n-apps\endcsname{12}
\expandafter\def\csname pct@rev.desc.d1.sspro.p2.T.grid2-d3.n-apps\endcsname{1200.0}
\expandafter\def\csname ppt@rev.desc.d1.sspro.p2.T.grid2-d3.n-apps\endcsname{$+$1200.0}
\expandafter\def\csname val@rev.desc.d1.sspro.p2.T.grid2-d3.n-tasks\endcsname{96}
\expandafter\def\csname cnt@rev.desc.d1.sspro.p2.T.grid2-d3.n-tasks\endcsname{96}
\expandafter\def\csname pct@rev.desc.d1.sspro.p2.T.grid2-d3.n-tasks\endcsname{9600.0}
\expandafter\def\csname ppt@rev.desc.d1.sspro.p2.T.grid2-d3.n-tasks\endcsname{$+$9600.0}
\expandafter\def\csname val@rev.desc.d1.sspro.p2.T.grid2-d3.sign-differs\endcsname{0}
\expandafter\def\csname cnt@rev.desc.d1.sspro.p2.T.grid2-d3.sign-differs\endcsname{0}
\expandafter\def\csname pct@rev.desc.d1.sspro.p2.T.grid2-d3.sign-differs\endcsname{0.0}
\expandafter\def\csname ppt@rev.desc.d1.sspro.p2.T.grid2-d3.sign-differs\endcsname{0.0}
\expandafter\def\csname val@rev.desc.d1.sspro.p2.T.grid2-d3.tau-source.ci90.0\endcsname{0.021}
\expandafter\def\csname cnt@rev.desc.d1.sspro.p2.T.grid2-d3.tau-source.ci90.0\endcsname{0}
\expandafter\def\csname pct@rev.desc.d1.sspro.p2.T.grid2-d3.tau-source.ci90.0\endcsname{2.1}
\expandafter\def\csname ppt@rev.desc.d1.sspro.p2.T.grid2-d3.tau-source.ci90.0\endcsname{$+$2.1}
\expandafter\def\csname val@rev.desc.d1.sspro.p2.T.grid2-d3.tau-source.ci90.1\endcsname{0.219}
\expandafter\def\csname cnt@rev.desc.d1.sspro.p2.T.grid2-d3.tau-source.ci90.1\endcsname{0}
\expandafter\def\csname pct@rev.desc.d1.sspro.p2.T.grid2-d3.tau-source.ci90.1\endcsname{21.9}
\expandafter\def\csname ppt@rev.desc.d1.sspro.p2.T.grid2-d3.tau-source.ci90.1\endcsname{$+$21.9}
\expandafter\def\csname val@rev.desc.d1.sspro.p2.T.grid2-d3.tau-source.ci95-excludes-0\endcsname{1}
\expandafter\def\csname cnt@rev.desc.d1.sspro.p2.T.grid2-d3.tau-source.ci95-excludes-0\endcsname{1}
\expandafter\def\csname pct@rev.desc.d1.sspro.p2.T.grid2-d3.tau-source.ci95-excludes-0\endcsname{100.0}
\expandafter\def\csname ppt@rev.desc.d1.sspro.p2.T.grid2-d3.tau-source.ci95-excludes-0\endcsname{$+$100.0}
\expandafter\def\csname val@rev.desc.d1.sspro.p2.T.grid2-d3.tau-source.ci95.0\endcsname{0.010}
\expandafter\def\csname cnt@rev.desc.d1.sspro.p2.T.grid2-d3.tau-source.ci95.0\endcsname{0}
\expandafter\def\csname pct@rev.desc.d1.sspro.p2.T.grid2-d3.tau-source.ci95.0\endcsname{1.0}
\expandafter\def\csname ppt@rev.desc.d1.sspro.p2.T.grid2-d3.tau-source.ci95.0\endcsname{$+$1.0}
\expandafter\def\csname val@rev.desc.d1.sspro.p2.T.grid2-d3.tau-source.ci95.1\endcsname{0.229}
\expandafter\def\csname cnt@rev.desc.d1.sspro.p2.T.grid2-d3.tau-source.ci95.1\endcsname{0}
\expandafter\def\csname pct@rev.desc.d1.sspro.p2.T.grid2-d3.tau-source.ci95.1\endcsname{22.9}
\expandafter\def\csname ppt@rev.desc.d1.sspro.p2.T.grid2-d3.tau-source.ci95.1\endcsname{$+$22.9}
\expandafter\def\csname val@rev.desc.d1.sspro.p2.T.grid2-d3.tau-source.mean\endcsname{0.115}
\expandafter\def\csname cnt@rev.desc.d1.sspro.p2.T.grid2-d3.tau-source.mean\endcsname{0}
\expandafter\def\csname pct@rev.desc.d1.sspro.p2.T.grid2-d3.tau-source.mean\endcsname{11.5}
\expandafter\def\csname ppt@rev.desc.d1.sspro.p2.T.grid2-d3.tau-source.mean\endcsname{$+$11.5}
\expandafter\def\csname val@rev.desc.d1.sspro.p2.T.grid2-d3.tau-source.n-apps\endcsname{12}
\expandafter\def\csname cnt@rev.desc.d1.sspro.p2.T.grid2-d3.tau-source.n-apps\endcsname{12}
\expandafter\def\csname pct@rev.desc.d1.sspro.p2.T.grid2-d3.tau-source.n-apps\endcsname{1200.0}
\expandafter\def\csname ppt@rev.desc.d1.sspro.p2.T.grid2-d3.tau-source.n-apps\endcsname{$+$1200.0}
\expandafter\def\csname val@rev.desc.d1.sspro.p2.T.grid2-d3.tau-source.tost1.ci90-inside\endcsname{0}
\expandafter\def\csname cnt@rev.desc.d1.sspro.p2.T.grid2-d3.tau-source.tost1.ci90-inside\endcsname{0}
\expandafter\def\csname pct@rev.desc.d1.sspro.p2.T.grid2-d3.tau-source.tost1.ci90-inside\endcsname{0.0}
\expandafter\def\csname ppt@rev.desc.d1.sspro.p2.T.grid2-d3.tau-source.tost1.ci90-inside\endcsname{0.0}
\expandafter\def\csname val@rev.desc.d1.sspro.p2.T.grid2-d3.tau-source.tost1.equivalent\endcsname{0}
\expandafter\def\csname cnt@rev.desc.d1.sspro.p2.T.grid2-d3.tau-source.tost1.equivalent\endcsname{0}
\expandafter\def\csname pct@rev.desc.d1.sspro.p2.T.grid2-d3.tau-source.tost1.equivalent\endcsname{0.0}
\expandafter\def\csname ppt@rev.desc.d1.sspro.p2.T.grid2-d3.tau-source.tost1.equivalent\endcsname{0.0}
\expandafter\def\csname val@rev.desc.d1.sspro.p2.T.grid2-d3.tau-source.tost1.p\endcsname{0.975}
\expandafter\def\csname cnt@rev.desc.d1.sspro.p2.T.grid2-d3.tau-source.tost1.p\endcsname{1}
\expandafter\def\csname pct@rev.desc.d1.sspro.p2.T.grid2-d3.tau-source.tost1.p\endcsname{97.5}
\expandafter\def\csname ppt@rev.desc.d1.sspro.p2.T.grid2-d3.tau-source.tost1.p\endcsname{$+$97.5}
\expandafter\def\csname val@rev.desc.d1.sspro.p2.T.grid2-d3.tau-source.tost5.ci90-inside\endcsname{0}
\expandafter\def\csname cnt@rev.desc.d1.sspro.p2.T.grid2-d3.tau-source.tost5.ci90-inside\endcsname{0}
\expandafter\def\csname pct@rev.desc.d1.sspro.p2.T.grid2-d3.tau-source.tost5.ci90-inside\endcsname{0.0}
\expandafter\def\csname ppt@rev.desc.d1.sspro.p2.T.grid2-d3.tau-source.tost5.ci90-inside\endcsname{0.0}
\expandafter\def\csname val@rev.desc.d1.sspro.p2.T.grid2-d3.tau-source.tost5.equivalent\endcsname{0}
\expandafter\def\csname cnt@rev.desc.d1.sspro.p2.T.grid2-d3.tau-source.tost5.equivalent\endcsname{0}
\expandafter\def\csname pct@rev.desc.d1.sspro.p2.T.grid2-d3.tau-source.tost5.equivalent\endcsname{0.0}
\expandafter\def\csname ppt@rev.desc.d1.sspro.p2.T.grid2-d3.tau-source.tost5.equivalent\endcsname{0.0}
\expandafter\def\csname val@rev.desc.d1.sspro.p2.T.grid2-d3.tau-source.tost5.p\endcsname{0.867}
\expandafter\def\csname cnt@rev.desc.d1.sspro.p2.T.grid2-d3.tau-source.tost5.p\endcsname{1}
\expandafter\def\csname pct@rev.desc.d1.sspro.p2.T.grid2-d3.tau-source.tost5.p\endcsname{86.7}
\expandafter\def\csname ppt@rev.desc.d1.sspro.p2.T.grid2-d3.tau-source.tost5.p\endcsname{$+$86.7}
\expandafter\def\csname val@rev.desc.d1.sspro.p2.T.grid2-d3.tau-target.ci90.0\endcsname{0.073}
\expandafter\def\csname cnt@rev.desc.d1.sspro.p2.T.grid2-d3.tau-target.ci90.0\endcsname{0}
\expandafter\def\csname pct@rev.desc.d1.sspro.p2.T.grid2-d3.tau-target.ci90.0\endcsname{7.3}
\expandafter\def\csname ppt@rev.desc.d1.sspro.p2.T.grid2-d3.tau-target.ci90.0\endcsname{$+$7.3}
\expandafter\def\csname val@rev.desc.d1.sspro.p2.T.grid2-d3.tau-target.ci90.1\endcsname{0.219}
\expandafter\def\csname cnt@rev.desc.d1.sspro.p2.T.grid2-d3.tau-target.ci90.1\endcsname{0}
\expandafter\def\csname pct@rev.desc.d1.sspro.p2.T.grid2-d3.tau-target.ci90.1\endcsname{21.9}
\expandafter\def\csname ppt@rev.desc.d1.sspro.p2.T.grid2-d3.tau-target.ci90.1\endcsname{$+$21.9}
\expandafter\def\csname val@rev.desc.d1.sspro.p2.T.grid2-d3.tau-target.ci95-excludes-0\endcsname{1}
\expandafter\def\csname cnt@rev.desc.d1.sspro.p2.T.grid2-d3.tau-target.ci95-excludes-0\endcsname{1}
\expandafter\def\csname pct@rev.desc.d1.sspro.p2.T.grid2-d3.tau-target.ci95-excludes-0\endcsname{100.0}
\expandafter\def\csname ppt@rev.desc.d1.sspro.p2.T.grid2-d3.tau-target.ci95-excludes-0\endcsname{$+$100.0}
\expandafter\def\csname val@rev.desc.d1.sspro.p2.T.grid2-d3.tau-target.ci95.0\endcsname{0.062}
\expandafter\def\csname cnt@rev.desc.d1.sspro.p2.T.grid2-d3.tau-target.ci95.0\endcsname{0}
\expandafter\def\csname pct@rev.desc.d1.sspro.p2.T.grid2-d3.tau-target.ci95.0\endcsname{6.2}
\expandafter\def\csname ppt@rev.desc.d1.sspro.p2.T.grid2-d3.tau-target.ci95.0\endcsname{$+$6.2}
\expandafter\def\csname val@rev.desc.d1.sspro.p2.T.grid2-d3.tau-target.ci95.1\endcsname{0.240}
\expandafter\def\csname cnt@rev.desc.d1.sspro.p2.T.grid2-d3.tau-target.ci95.1\endcsname{0}
\expandafter\def\csname pct@rev.desc.d1.sspro.p2.T.grid2-d3.tau-target.ci95.1\endcsname{24.0}
\expandafter\def\csname ppt@rev.desc.d1.sspro.p2.T.grid2-d3.tau-target.ci95.1\endcsname{$+$24.0}
\expandafter\def\csname val@rev.desc.d1.sspro.p2.T.grid2-d3.tau-target.mean\endcsname{0.146}
\expandafter\def\csname cnt@rev.desc.d1.sspro.p2.T.grid2-d3.tau-target.mean\endcsname{0}
\expandafter\def\csname pct@rev.desc.d1.sspro.p2.T.grid2-d3.tau-target.mean\endcsname{14.6}
\expandafter\def\csname ppt@rev.desc.d1.sspro.p2.T.grid2-d3.tau-target.mean\endcsname{$+$14.6}
\expandafter\def\csname val@rev.desc.d1.sspro.p2.T.grid2-d3.tau-target.n-apps\endcsname{12}
\expandafter\def\csname cnt@rev.desc.d1.sspro.p2.T.grid2-d3.tau-target.n-apps\endcsname{12}
\expandafter\def\csname pct@rev.desc.d1.sspro.p2.T.grid2-d3.tau-target.n-apps\endcsname{1200.0}
\expandafter\def\csname ppt@rev.desc.d1.sspro.p2.T.grid2-d3.tau-target.n-apps\endcsname{$+$1200.0}
\expandafter\def\csname val@rev.desc.d1.sspro.p2.T.grid2-d3.tau-target.tost1.ci90-inside\endcsname{0}
\expandafter\def\csname cnt@rev.desc.d1.sspro.p2.T.grid2-d3.tau-target.tost1.ci90-inside\endcsname{0}
\expandafter\def\csname pct@rev.desc.d1.sspro.p2.T.grid2-d3.tau-target.tost1.ci90-inside\endcsname{0.0}
\expandafter\def\csname ppt@rev.desc.d1.sspro.p2.T.grid2-d3.tau-target.tost1.ci90-inside\endcsname{0.0}
\expandafter\def\csname val@rev.desc.d1.sspro.p2.T.grid2-d3.tau-target.tost1.equivalent\endcsname{0}
\expandafter\def\csname cnt@rev.desc.d1.sspro.p2.T.grid2-d3.tau-target.tost1.equivalent\endcsname{0}
\expandafter\def\csname pct@rev.desc.d1.sspro.p2.T.grid2-d3.tau-target.tost1.equivalent\endcsname{0.0}
\expandafter\def\csname ppt@rev.desc.d1.sspro.p2.T.grid2-d3.tau-target.tost1.equivalent\endcsname{0.0}
\expandafter\def\csname val@rev.desc.d1.sspro.p2.T.grid2-d3.tau-target.tost1.p\endcsname{1.000}
\expandafter\def\csname cnt@rev.desc.d1.sspro.p2.T.grid2-d3.tau-target.tost1.p\endcsname{1}
\expandafter\def\csname pct@rev.desc.d1.sspro.p2.T.grid2-d3.tau-target.tost1.p\endcsname{100.0}
\expandafter\def\csname ppt@rev.desc.d1.sspro.p2.T.grid2-d3.tau-target.tost1.p\endcsname{$+$100.0}
\expandafter\def\csname val@rev.desc.d1.sspro.p2.T.grid2-d3.tau-target.tost5.ci90-inside\endcsname{0}
\expandafter\def\csname cnt@rev.desc.d1.sspro.p2.T.grid2-d3.tau-target.tost5.ci90-inside\endcsname{0}
\expandafter\def\csname pct@rev.desc.d1.sspro.p2.T.grid2-d3.tau-target.tost5.ci90-inside\endcsname{0.0}
\expandafter\def\csname ppt@rev.desc.d1.sspro.p2.T.grid2-d3.tau-target.tost5.ci90-inside\endcsname{0.0}
\expandafter\def\csname val@rev.desc.d1.sspro.p2.T.grid2-d3.tau-target.tost5.equivalent\endcsname{0}
\expandafter\def\csname cnt@rev.desc.d1.sspro.p2.T.grid2-d3.tau-target.tost5.equivalent\endcsname{0}
\expandafter\def\csname pct@rev.desc.d1.sspro.p2.T.grid2-d3.tau-target.tost5.equivalent\endcsname{0.0}
\expandafter\def\csname ppt@rev.desc.d1.sspro.p2.T.grid2-d3.tau-target.tost5.equivalent\endcsname{0.0}
\expandafter\def\csname val@rev.desc.d1.sspro.p2.T.grid2-d3.tau-target.tost5.p\endcsname{0.992}
\expandafter\def\csname cnt@rev.desc.d1.sspro.p2.T.grid2-d3.tau-target.tost5.p\endcsname{1}
\expandafter\def\csname pct@rev.desc.d1.sspro.p2.T.grid2-d3.tau-target.tost5.p\endcsname{99.2}
\expandafter\def\csname ppt@rev.desc.d1.sspro.p2.T.grid2-d3.tau-target.tost5.p\endcsname{$+$99.2}
\expandafter\def\csname val@rev.desc.d1.sspro.p2.T.grid2-d3.zero-estimate\endcsname{0}
\expandafter\def\csname cnt@rev.desc.d1.sspro.p2.T.grid2-d3.zero-estimate\endcsname{0}
\expandafter\def\csname pct@rev.desc.d1.sspro.p2.T.grid2-d3.zero-estimate\endcsname{0.0}
\expandafter\def\csname ppt@rev.desc.d1.sspro.p2.T.grid2-d3.zero-estimate\endcsname{0.0}
\expandafter\def\csname val@rev.desc.d1.sspro.p2.T.zoomclick-d3.delta-tau.ci90.0\endcsname{-0.062}
\expandafter\def\csname cnt@rev.desc.d1.sspro.p2.T.zoomclick-d3.delta-tau.ci90.0\endcsname{0}
\expandafter\def\csname pct@rev.desc.d1.sspro.p2.T.zoomclick-d3.delta-tau.ci90.0\endcsname{$-$6.2}
\expandafter\def\csname ppt@rev.desc.d1.sspro.p2.T.zoomclick-d3.delta-tau.ci90.0\endcsname{$-$6.2}
\expandafter\def\csname val@rev.desc.d1.sspro.p2.T.zoomclick-d3.delta-tau.ci90.1\endcsname{0.042}
\expandafter\def\csname cnt@rev.desc.d1.sspro.p2.T.zoomclick-d3.delta-tau.ci90.1\endcsname{0}
\expandafter\def\csname pct@rev.desc.d1.sspro.p2.T.zoomclick-d3.delta-tau.ci90.1\endcsname{4.2}
\expandafter\def\csname ppt@rev.desc.d1.sspro.p2.T.zoomclick-d3.delta-tau.ci90.1\endcsname{$+$4.2}
\expandafter\def\csname val@rev.desc.d1.sspro.p2.T.zoomclick-d3.delta-tau.ci95-excludes-0\endcsname{0}
\expandafter\def\csname cnt@rev.desc.d1.sspro.p2.T.zoomclick-d3.delta-tau.ci95-excludes-0\endcsname{0}
\expandafter\def\csname pct@rev.desc.d1.sspro.p2.T.zoomclick-d3.delta-tau.ci95-excludes-0\endcsname{0.0}
\expandafter\def\csname ppt@rev.desc.d1.sspro.p2.T.zoomclick-d3.delta-tau.ci95-excludes-0\endcsname{0.0}
\expandafter\def\csname val@rev.desc.d1.sspro.p2.T.zoomclick-d3.delta-tau.ci95.0\endcsname{-0.073}
\expandafter\def\csname cnt@rev.desc.d1.sspro.p2.T.zoomclick-d3.delta-tau.ci95.0\endcsname{0}
\expandafter\def\csname pct@rev.desc.d1.sspro.p2.T.zoomclick-d3.delta-tau.ci95.0\endcsname{$-$7.3}
\expandafter\def\csname ppt@rev.desc.d1.sspro.p2.T.zoomclick-d3.delta-tau.ci95.0\endcsname{$-$7.3}
\expandafter\def\csname val@rev.desc.d1.sspro.p2.T.zoomclick-d3.delta-tau.ci95.1\endcsname{0.052}
\expandafter\def\csname cnt@rev.desc.d1.sspro.p2.T.zoomclick-d3.delta-tau.ci95.1\endcsname{0}
\expandafter\def\csname pct@rev.desc.d1.sspro.p2.T.zoomclick-d3.delta-tau.ci95.1\endcsname{5.2}
\expandafter\def\csname ppt@rev.desc.d1.sspro.p2.T.zoomclick-d3.delta-tau.ci95.1\endcsname{$+$5.2}
\expandafter\def\csname val@rev.desc.d1.sspro.p2.T.zoomclick-d3.delta-tau.mean\endcsname{-0.010}
\expandafter\def\csname cnt@rev.desc.d1.sspro.p2.T.zoomclick-d3.delta-tau.mean\endcsname{0}
\expandafter\def\csname pct@rev.desc.d1.sspro.p2.T.zoomclick-d3.delta-tau.mean\endcsname{$-$1.0}
\expandafter\def\csname ppt@rev.desc.d1.sspro.p2.T.zoomclick-d3.delta-tau.mean\endcsname{$-$1.0}
\expandafter\def\csname val@rev.desc.d1.sspro.p2.T.zoomclick-d3.delta-tau.n-apps\endcsname{12}
\expandafter\def\csname cnt@rev.desc.d1.sspro.p2.T.zoomclick-d3.delta-tau.n-apps\endcsname{12}
\expandafter\def\csname pct@rev.desc.d1.sspro.p2.T.zoomclick-d3.delta-tau.n-apps\endcsname{1200.0}
\expandafter\def\csname ppt@rev.desc.d1.sspro.p2.T.zoomclick-d3.delta-tau.n-apps\endcsname{$+$1200.0}
\expandafter\def\csname val@rev.desc.d1.sspro.p2.T.zoomclick-d3.delta-tau.tost1.ci90-inside\endcsname{0}
\expandafter\def\csname cnt@rev.desc.d1.sspro.p2.T.zoomclick-d3.delta-tau.tost1.ci90-inside\endcsname{0}
\expandafter\def\csname pct@rev.desc.d1.sspro.p2.T.zoomclick-d3.delta-tau.tost1.ci90-inside\endcsname{0.0}
\expandafter\def\csname ppt@rev.desc.d1.sspro.p2.T.zoomclick-d3.delta-tau.tost1.ci90-inside\endcsname{0.0}
\expandafter\def\csname val@rev.desc.d1.sspro.p2.T.zoomclick-d3.delta-tau.tost1.equivalent\endcsname{0}
\expandafter\def\csname cnt@rev.desc.d1.sspro.p2.T.zoomclick-d3.delta-tau.tost1.equivalent\endcsname{0}
\expandafter\def\csname pct@rev.desc.d1.sspro.p2.T.zoomclick-d3.delta-tau.tost1.equivalent\endcsname{0.0}
\expandafter\def\csname ppt@rev.desc.d1.sspro.p2.T.zoomclick-d3.delta-tau.tost1.equivalent\endcsname{0.0}
\expandafter\def\csname val@rev.desc.d1.sspro.p2.T.zoomclick-d3.delta-tau.tost1.p\endcsname{0.554}
\expandafter\def\csname cnt@rev.desc.d1.sspro.p2.T.zoomclick-d3.delta-tau.tost1.p\endcsname{1}
\expandafter\def\csname pct@rev.desc.d1.sspro.p2.T.zoomclick-d3.delta-tau.tost1.p\endcsname{55.4}
\expandafter\def\csname ppt@rev.desc.d1.sspro.p2.T.zoomclick-d3.delta-tau.tost1.p\endcsname{$+$55.4}
\expandafter\def\csname val@rev.desc.d1.sspro.p2.T.zoomclick-d3.delta-tau.tost5.ci90-inside\endcsname{0}
\expandafter\def\csname cnt@rev.desc.d1.sspro.p2.T.zoomclick-d3.delta-tau.tost5.ci90-inside\endcsname{0}
\expandafter\def\csname pct@rev.desc.d1.sspro.p2.T.zoomclick-d3.delta-tau.tost5.ci90-inside\endcsname{0.0}
\expandafter\def\csname ppt@rev.desc.d1.sspro.p2.T.zoomclick-d3.delta-tau.tost5.ci90-inside\endcsname{0.0}
\expandafter\def\csname val@rev.desc.d1.sspro.p2.T.zoomclick-d3.delta-tau.tost5.equivalent\endcsname{0}
\expandafter\def\csname cnt@rev.desc.d1.sspro.p2.T.zoomclick-d3.delta-tau.tost5.equivalent\endcsname{0}
\expandafter\def\csname pct@rev.desc.d1.sspro.p2.T.zoomclick-d3.delta-tau.tost5.equivalent\endcsname{0.0}
\expandafter\def\csname ppt@rev.desc.d1.sspro.p2.T.zoomclick-d3.delta-tau.tost5.equivalent\endcsname{0.0}
\expandafter\def\csname val@rev.desc.d1.sspro.p2.T.zoomclick-d3.delta-tau.tost5.p\endcsname{0.122}
\expandafter\def\csname cnt@rev.desc.d1.sspro.p2.T.zoomclick-d3.delta-tau.tost5.p\endcsname{0}
\expandafter\def\csname pct@rev.desc.d1.sspro.p2.T.zoomclick-d3.delta-tau.tost5.p\endcsname{12.2}
\expandafter\def\csname ppt@rev.desc.d1.sspro.p2.T.zoomclick-d3.delta-tau.tost5.p\endcsname{$+$12.2}
\expandafter\def\csname val@rev.desc.d1.sspro.p2.T.zoomclick-d3.material-flip\endcsname{0}
\expandafter\def\csname cnt@rev.desc.d1.sspro.p2.T.zoomclick-d3.material-flip\endcsname{0}
\expandafter\def\csname pct@rev.desc.d1.sspro.p2.T.zoomclick-d3.material-flip\endcsname{0.0}
\expandafter\def\csname ppt@rev.desc.d1.sspro.p2.T.zoomclick-d3.material-flip\endcsname{0.0}
\expandafter\def\csname val@rev.desc.d1.sspro.p2.T.zoomclick-d3.n-apps\endcsname{12}
\expandafter\def\csname cnt@rev.desc.d1.sspro.p2.T.zoomclick-d3.n-apps\endcsname{12}
\expandafter\def\csname pct@rev.desc.d1.sspro.p2.T.zoomclick-d3.n-apps\endcsname{1200.0}
\expandafter\def\csname ppt@rev.desc.d1.sspro.p2.T.zoomclick-d3.n-apps\endcsname{$+$1200.0}
\expandafter\def\csname val@rev.desc.d1.sspro.p2.T.zoomclick-d3.n-tasks\endcsname{96}
\expandafter\def\csname cnt@rev.desc.d1.sspro.p2.T.zoomclick-d3.n-tasks\endcsname{96}
\expandafter\def\csname pct@rev.desc.d1.sspro.p2.T.zoomclick-d3.n-tasks\endcsname{9600.0}
\expandafter\def\csname ppt@rev.desc.d1.sspro.p2.T.zoomclick-d3.n-tasks\endcsname{$+$9600.0}
\expandafter\def\csname val@rev.desc.d1.sspro.p2.T.zoomclick-d3.sign-differs\endcsname{0}
\expandafter\def\csname cnt@rev.desc.d1.sspro.p2.T.zoomclick-d3.sign-differs\endcsname{0}
\expandafter\def\csname pct@rev.desc.d1.sspro.p2.T.zoomclick-d3.sign-differs\endcsname{0.0}
\expandafter\def\csname ppt@rev.desc.d1.sspro.p2.T.zoomclick-d3.sign-differs\endcsname{0.0}
\expandafter\def\csname val@rev.desc.d1.sspro.p2.T.zoomclick-d3.tau-source.ci90.0\endcsname{0.104}
\expandafter\def\csname cnt@rev.desc.d1.sspro.p2.T.zoomclick-d3.tau-source.ci90.0\endcsname{0}
\expandafter\def\csname pct@rev.desc.d1.sspro.p2.T.zoomclick-d3.tau-source.ci90.0\endcsname{10.4}
\expandafter\def\csname ppt@rev.desc.d1.sspro.p2.T.zoomclick-d3.tau-source.ci90.0\endcsname{$+$10.4}
\expandafter\def\csname val@rev.desc.d1.sspro.p2.T.zoomclick-d3.tau-source.ci90.1\endcsname{0.271}
\expandafter\def\csname cnt@rev.desc.d1.sspro.p2.T.zoomclick-d3.tau-source.ci90.1\endcsname{0}
\expandafter\def\csname pct@rev.desc.d1.sspro.p2.T.zoomclick-d3.tau-source.ci90.1\endcsname{27.1}
\expandafter\def\csname ppt@rev.desc.d1.sspro.p2.T.zoomclick-d3.tau-source.ci90.1\endcsname{$+$27.1}
\expandafter\def\csname val@rev.desc.d1.sspro.p2.T.zoomclick-d3.tau-source.ci95-excludes-0\endcsname{1}
\expandafter\def\csname cnt@rev.desc.d1.sspro.p2.T.zoomclick-d3.tau-source.ci95-excludes-0\endcsname{1}
\expandafter\def\csname pct@rev.desc.d1.sspro.p2.T.zoomclick-d3.tau-source.ci95-excludes-0\endcsname{100.0}
\expandafter\def\csname ppt@rev.desc.d1.sspro.p2.T.zoomclick-d3.tau-source.ci95-excludes-0\endcsname{$+$100.0}
\expandafter\def\csname val@rev.desc.d1.sspro.p2.T.zoomclick-d3.tau-source.ci95.0\endcsname{0.094}
\expandafter\def\csname cnt@rev.desc.d1.sspro.p2.T.zoomclick-d3.tau-source.ci95.0\endcsname{0}
\expandafter\def\csname pct@rev.desc.d1.sspro.p2.T.zoomclick-d3.tau-source.ci95.0\endcsname{9.4}
\expandafter\def\csname ppt@rev.desc.d1.sspro.p2.T.zoomclick-d3.tau-source.ci95.0\endcsname{$+$9.4}
\expandafter\def\csname val@rev.desc.d1.sspro.p2.T.zoomclick-d3.tau-source.ci95.1\endcsname{0.292}
\expandafter\def\csname cnt@rev.desc.d1.sspro.p2.T.zoomclick-d3.tau-source.ci95.1\endcsname{0}
\expandafter\def\csname pct@rev.desc.d1.sspro.p2.T.zoomclick-d3.tau-source.ci95.1\endcsname{29.2}
\expandafter\def\csname ppt@rev.desc.d1.sspro.p2.T.zoomclick-d3.tau-source.ci95.1\endcsname{$+$29.2}
\expandafter\def\csname val@rev.desc.d1.sspro.p2.T.zoomclick-d3.tau-source.mean\endcsname{0.177}
\expandafter\def\csname cnt@rev.desc.d1.sspro.p2.T.zoomclick-d3.tau-source.mean\endcsname{0}
\expandafter\def\csname pct@rev.desc.d1.sspro.p2.T.zoomclick-d3.tau-source.mean\endcsname{17.7}
\expandafter\def\csname ppt@rev.desc.d1.sspro.p2.T.zoomclick-d3.tau-source.mean\endcsname{$+$17.7}
\expandafter\def\csname val@rev.desc.d1.sspro.p2.T.zoomclick-d3.tau-source.n-apps\endcsname{12}
\expandafter\def\csname cnt@rev.desc.d1.sspro.p2.T.zoomclick-d3.tau-source.n-apps\endcsname{12}
\expandafter\def\csname pct@rev.desc.d1.sspro.p2.T.zoomclick-d3.tau-source.n-apps\endcsname{1200.0}
\expandafter\def\csname ppt@rev.desc.d1.sspro.p2.T.zoomclick-d3.tau-source.n-apps\endcsname{$+$1200.0}
\expandafter\def\csname val@rev.desc.d1.sspro.p2.T.zoomclick-d3.tau-source.tost1.ci90-inside\endcsname{0}
\expandafter\def\csname cnt@rev.desc.d1.sspro.p2.T.zoomclick-d3.tau-source.tost1.ci90-inside\endcsname{0}
\expandafter\def\csname pct@rev.desc.d1.sspro.p2.T.zoomclick-d3.tau-source.tost1.ci90-inside\endcsname{0.0}
\expandafter\def\csname ppt@rev.desc.d1.sspro.p2.T.zoomclick-d3.tau-source.tost1.ci90-inside\endcsname{0.0}
\expandafter\def\csname val@rev.desc.d1.sspro.p2.T.zoomclick-d3.tau-source.tost1.equivalent\endcsname{0}
\expandafter\def\csname cnt@rev.desc.d1.sspro.p2.T.zoomclick-d3.tau-source.tost1.equivalent\endcsname{0}
\expandafter\def\csname pct@rev.desc.d1.sspro.p2.T.zoomclick-d3.tau-source.tost1.equivalent\endcsname{0.0}
\expandafter\def\csname ppt@rev.desc.d1.sspro.p2.T.zoomclick-d3.tau-source.tost1.equivalent\endcsname{0.0}
\expandafter\def\csname val@rev.desc.d1.sspro.p2.T.zoomclick-d3.tau-source.tost1.p\endcsname{1.000}
\expandafter\def\csname cnt@rev.desc.d1.sspro.p2.T.zoomclick-d3.tau-source.tost1.p\endcsname{1}
\expandafter\def\csname pct@rev.desc.d1.sspro.p2.T.zoomclick-d3.tau-source.tost1.p\endcsname{100.0}
\expandafter\def\csname ppt@rev.desc.d1.sspro.p2.T.zoomclick-d3.tau-source.tost1.p\endcsname{$+$100.0}
\expandafter\def\csname val@rev.desc.d1.sspro.p2.T.zoomclick-d3.tau-source.tost5.ci90-inside\endcsname{0}
\expandafter\def\csname cnt@rev.desc.d1.sspro.p2.T.zoomclick-d3.tau-source.tost5.ci90-inside\endcsname{0}
\expandafter\def\csname pct@rev.desc.d1.sspro.p2.T.zoomclick-d3.tau-source.tost5.ci90-inside\endcsname{0.0}
\expandafter\def\csname ppt@rev.desc.d1.sspro.p2.T.zoomclick-d3.tau-source.tost5.ci90-inside\endcsname{0.0}
\expandafter\def\csname val@rev.desc.d1.sspro.p2.T.zoomclick-d3.tau-source.tost5.equivalent\endcsname{0}
\expandafter\def\csname cnt@rev.desc.d1.sspro.p2.T.zoomclick-d3.tau-source.tost5.equivalent\endcsname{0}
\expandafter\def\csname pct@rev.desc.d1.sspro.p2.T.zoomclick-d3.tau-source.tost5.equivalent\endcsname{0.0}
\expandafter\def\csname ppt@rev.desc.d1.sspro.p2.T.zoomclick-d3.tau-source.tost5.equivalent\endcsname{0.0}
\expandafter\def\csname val@rev.desc.d1.sspro.p2.T.zoomclick-d3.tau-source.tost5.p\endcsname{0.999}
\expandafter\def\csname cnt@rev.desc.d1.sspro.p2.T.zoomclick-d3.tau-source.tost5.p\endcsname{1}
\expandafter\def\csname pct@rev.desc.d1.sspro.p2.T.zoomclick-d3.tau-source.tost5.p\endcsname{99.9}
\expandafter\def\csname ppt@rev.desc.d1.sspro.p2.T.zoomclick-d3.tau-source.tost5.p\endcsname{$+$99.9}
\expandafter\def\csname val@rev.desc.d1.sspro.p2.T.zoomclick-d3.tau-target.ci90.0\endcsname{0.115}
\expandafter\def\csname cnt@rev.desc.d1.sspro.p2.T.zoomclick-d3.tau-target.ci90.0\endcsname{0}
\expandafter\def\csname pct@rev.desc.d1.sspro.p2.T.zoomclick-d3.tau-target.ci90.0\endcsname{11.5}
\expandafter\def\csname ppt@rev.desc.d1.sspro.p2.T.zoomclick-d3.tau-target.ci90.0\endcsname{$+$11.5}
\expandafter\def\csname val@rev.desc.d1.sspro.p2.T.zoomclick-d3.tau-target.ci90.1\endcsname{0.229}
\expandafter\def\csname cnt@rev.desc.d1.sspro.p2.T.zoomclick-d3.tau-target.ci90.1\endcsname{0}
\expandafter\def\csname pct@rev.desc.d1.sspro.p2.T.zoomclick-d3.tau-target.ci90.1\endcsname{22.9}
\expandafter\def\csname ppt@rev.desc.d1.sspro.p2.T.zoomclick-d3.tau-target.ci90.1\endcsname{$+$22.9}
\expandafter\def\csname val@rev.desc.d1.sspro.p2.T.zoomclick-d3.tau-target.ci95-excludes-0\endcsname{1}
\expandafter\def\csname cnt@rev.desc.d1.sspro.p2.T.zoomclick-d3.tau-target.ci95-excludes-0\endcsname{1}
\expandafter\def\csname pct@rev.desc.d1.sspro.p2.T.zoomclick-d3.tau-target.ci95-excludes-0\endcsname{100.0}
\expandafter\def\csname ppt@rev.desc.d1.sspro.p2.T.zoomclick-d3.tau-target.ci95-excludes-0\endcsname{$+$100.0}
\expandafter\def\csname val@rev.desc.d1.sspro.p2.T.zoomclick-d3.tau-target.ci95.0\endcsname{0.104}
\expandafter\def\csname cnt@rev.desc.d1.sspro.p2.T.zoomclick-d3.tau-target.ci95.0\endcsname{0}
\expandafter\def\csname pct@rev.desc.d1.sspro.p2.T.zoomclick-d3.tau-target.ci95.0\endcsname{10.4}
\expandafter\def\csname ppt@rev.desc.d1.sspro.p2.T.zoomclick-d3.tau-target.ci95.0\endcsname{$+$10.4}
\expandafter\def\csname val@rev.desc.d1.sspro.p2.T.zoomclick-d3.tau-target.ci95.1\endcsname{0.240}
\expandafter\def\csname cnt@rev.desc.d1.sspro.p2.T.zoomclick-d3.tau-target.ci95.1\endcsname{0}
\expandafter\def\csname pct@rev.desc.d1.sspro.p2.T.zoomclick-d3.tau-target.ci95.1\endcsname{24.0}
\expandafter\def\csname ppt@rev.desc.d1.sspro.p2.T.zoomclick-d3.tau-target.ci95.1\endcsname{$+$24.0}
\expandafter\def\csname val@rev.desc.d1.sspro.p2.T.zoomclick-d3.tau-target.mean\endcsname{0.167}
\expandafter\def\csname cnt@rev.desc.d1.sspro.p2.T.zoomclick-d3.tau-target.mean\endcsname{0}
\expandafter\def\csname pct@rev.desc.d1.sspro.p2.T.zoomclick-d3.tau-target.mean\endcsname{16.7}
\expandafter\def\csname ppt@rev.desc.d1.sspro.p2.T.zoomclick-d3.tau-target.mean\endcsname{$+$16.7}
\expandafter\def\csname val@rev.desc.d1.sspro.p2.T.zoomclick-d3.tau-target.n-apps\endcsname{12}
\expandafter\def\csname cnt@rev.desc.d1.sspro.p2.T.zoomclick-d3.tau-target.n-apps\endcsname{12}
\expandafter\def\csname pct@rev.desc.d1.sspro.p2.T.zoomclick-d3.tau-target.n-apps\endcsname{1200.0}
\expandafter\def\csname ppt@rev.desc.d1.sspro.p2.T.zoomclick-d3.tau-target.n-apps\endcsname{$+$1200.0}
\expandafter\def\csname val@rev.desc.d1.sspro.p2.T.zoomclick-d3.tau-target.tost1.ci90-inside\endcsname{0}
\expandafter\def\csname cnt@rev.desc.d1.sspro.p2.T.zoomclick-d3.tau-target.tost1.ci90-inside\endcsname{0}
\expandafter\def\csname pct@rev.desc.d1.sspro.p2.T.zoomclick-d3.tau-target.tost1.ci90-inside\endcsname{0.0}
\expandafter\def\csname ppt@rev.desc.d1.sspro.p2.T.zoomclick-d3.tau-target.tost1.ci90-inside\endcsname{0.0}
\expandafter\def\csname val@rev.desc.d1.sspro.p2.T.zoomclick-d3.tau-target.tost1.equivalent\endcsname{0}
\expandafter\def\csname cnt@rev.desc.d1.sspro.p2.T.zoomclick-d3.tau-target.tost1.equivalent\endcsname{0}
\expandafter\def\csname pct@rev.desc.d1.sspro.p2.T.zoomclick-d3.tau-target.tost1.equivalent\endcsname{0.0}
\expandafter\def\csname ppt@rev.desc.d1.sspro.p2.T.zoomclick-d3.tau-target.tost1.equivalent\endcsname{0.0}
\expandafter\def\csname val@rev.desc.d1.sspro.p2.T.zoomclick-d3.tau-target.tost1.p\endcsname{1.000}
\expandafter\def\csname cnt@rev.desc.d1.sspro.p2.T.zoomclick-d3.tau-target.tost1.p\endcsname{1}
\expandafter\def\csname pct@rev.desc.d1.sspro.p2.T.zoomclick-d3.tau-target.tost1.p\endcsname{100.0}
\expandafter\def\csname ppt@rev.desc.d1.sspro.p2.T.zoomclick-d3.tau-target.tost1.p\endcsname{$+$100.0}
\expandafter\def\csname val@rev.desc.d1.sspro.p2.T.zoomclick-d3.tau-target.tost5.ci90-inside\endcsname{0}
\expandafter\def\csname cnt@rev.desc.d1.sspro.p2.T.zoomclick-d3.tau-target.tost5.ci90-inside\endcsname{0}
\expandafter\def\csname pct@rev.desc.d1.sspro.p2.T.zoomclick-d3.tau-target.tost5.ci90-inside\endcsname{0.0}
\expandafter\def\csname ppt@rev.desc.d1.sspro.p2.T.zoomclick-d3.tau-target.tost5.ci90-inside\endcsname{0.0}
\expandafter\def\csname val@rev.desc.d1.sspro.p2.T.zoomclick-d3.tau-target.tost5.equivalent\endcsname{0}
\expandafter\def\csname cnt@rev.desc.d1.sspro.p2.T.zoomclick-d3.tau-target.tost5.equivalent\endcsname{0}
\expandafter\def\csname pct@rev.desc.d1.sspro.p2.T.zoomclick-d3.tau-target.tost5.equivalent\endcsname{0.0}
\expandafter\def\csname ppt@rev.desc.d1.sspro.p2.T.zoomclick-d3.tau-target.tost5.equivalent\endcsname{0.0}
\expandafter\def\csname val@rev.desc.d1.sspro.p2.T.zoomclick-d3.tau-target.tost5.p\endcsname{1.000}
\expandafter\def\csname cnt@rev.desc.d1.sspro.p2.T.zoomclick-d3.tau-target.tost5.p\endcsname{1}
\expandafter\def\csname pct@rev.desc.d1.sspro.p2.T.zoomclick-d3.tau-target.tost5.p\endcsname{100.0}
\expandafter\def\csname ppt@rev.desc.d1.sspro.p2.T.zoomclick-d3.tau-target.tost5.p\endcsname{$+$100.0}
\expandafter\def\csname val@rev.desc.d1.sspro.p2.T.zoomclick-d3.zero-estimate\endcsname{0}
\expandafter\def\csname cnt@rev.desc.d1.sspro.p2.T.zoomclick-d3.zero-estimate\endcsname{0}
\expandafter\def\csname pct@rev.desc.d1.sspro.p2.T.zoomclick-d3.zero-estimate\endcsname{0.0}
\expandafter\def\csname ppt@rev.desc.d1.sspro.p2.T.zoomclick-d3.zero-estimate\endcsname{0.0}
\expandafter\def\csname val@rev.desc.d1.sspro.p2.Tsealed.center-d1.n-tasks\endcsname{96}
\expandafter\def\csname cnt@rev.desc.d1.sspro.p2.Tsealed.center-d1.n-tasks\endcsname{96}
\expandafter\def\csname pct@rev.desc.d1.sspro.p2.Tsealed.center-d1.n-tasks\endcsname{9600.0}
\expandafter\def\csname ppt@rev.desc.d1.sspro.p2.Tsealed.center-d1.n-tasks\endcsname{$+$9600.0}
\expandafter\def\csname val@rev.desc.d1.sspro.p2.Tsealed.center-d1.paper-tau-T\endcsname{0.156}
\expandafter\def\csname cnt@rev.desc.d1.sspro.p2.Tsealed.center-d1.paper-tau-T\endcsname{0}
\expandafter\def\csname pct@rev.desc.d1.sspro.p2.Tsealed.center-d1.paper-tau-T\endcsname{15.6}
\expandafter\def\csname ppt@rev.desc.d1.sspro.p2.Tsealed.center-d1.paper-tau-T\endcsname{$+$15.6}
\expandafter\def\csname val@rev.desc.d1.sspro.p2.Tsealed.center-d1.reproduces\endcsname{1}
\expandafter\def\csname cnt@rev.desc.d1.sspro.p2.Tsealed.center-d1.reproduces\endcsname{1}
\expandafter\def\csname pct@rev.desc.d1.sspro.p2.Tsealed.center-d1.reproduces\endcsname{100.0}
\expandafter\def\csname ppt@rev.desc.d1.sspro.p2.Tsealed.center-d1.reproduces\endcsname{$+$100.0}
\expandafter\def\csname val@rev.desc.d1.sspro.p2.Tsealed.center-d1.reproduces-ci\endcsname{1}
\expandafter\def\csname cnt@rev.desc.d1.sspro.p2.Tsealed.center-d1.reproduces-ci\endcsname{1}
\expandafter\def\csname pct@rev.desc.d1.sspro.p2.Tsealed.center-d1.reproduces-ci\endcsname{100.0}
\expandafter\def\csname ppt@rev.desc.d1.sspro.p2.Tsealed.center-d1.reproduces-ci\endcsname{$+$100.0}
\expandafter\def\csname val@rev.desc.d1.sspro.p2.Tsealed.center-d1.tau-target-T.ci90.0\endcsname{0.094}
\expandafter\def\csname cnt@rev.desc.d1.sspro.p2.Tsealed.center-d1.tau-target-T.ci90.0\endcsname{0}
\expandafter\def\csname pct@rev.desc.d1.sspro.p2.Tsealed.center-d1.tau-target-T.ci90.0\endcsname{9.4}
\expandafter\def\csname ppt@rev.desc.d1.sspro.p2.Tsealed.center-d1.tau-target-T.ci90.0\endcsname{$+$9.4}
\expandafter\def\csname val@rev.desc.d1.sspro.p2.Tsealed.center-d1.tau-target-T.ci90.1\endcsname{0.229}
\expandafter\def\csname cnt@rev.desc.d1.sspro.p2.Tsealed.center-d1.tau-target-T.ci90.1\endcsname{0}
\expandafter\def\csname pct@rev.desc.d1.sspro.p2.Tsealed.center-d1.tau-target-T.ci90.1\endcsname{22.9}
\expandafter\def\csname ppt@rev.desc.d1.sspro.p2.Tsealed.center-d1.tau-target-T.ci90.1\endcsname{$+$22.9}
\expandafter\def\csname val@rev.desc.d1.sspro.p2.Tsealed.center-d1.tau-target-T.ci95-excludes-0\endcsname{1}
\expandafter\def\csname cnt@rev.desc.d1.sspro.p2.Tsealed.center-d1.tau-target-T.ci95-excludes-0\endcsname{1}
\expandafter\def\csname pct@rev.desc.d1.sspro.p2.Tsealed.center-d1.tau-target-T.ci95-excludes-0\endcsname{100.0}
\expandafter\def\csname ppt@rev.desc.d1.sspro.p2.Tsealed.center-d1.tau-target-T.ci95-excludes-0\endcsname{$+$100.0}
\expandafter\def\csname val@rev.desc.d1.sspro.p2.Tsealed.center-d1.tau-target-T.ci95.0\endcsname{0.083}
\expandafter\def\csname cnt@rev.desc.d1.sspro.p2.Tsealed.center-d1.tau-target-T.ci95.0\endcsname{0}
\expandafter\def\csname pct@rev.desc.d1.sspro.p2.Tsealed.center-d1.tau-target-T.ci95.0\endcsname{8.3}
\expandafter\def\csname ppt@rev.desc.d1.sspro.p2.Tsealed.center-d1.tau-target-T.ci95.0\endcsname{$+$8.3}
\expandafter\def\csname val@rev.desc.d1.sspro.p2.Tsealed.center-d1.tau-target-T.ci95.1\endcsname{0.240}
\expandafter\def\csname cnt@rev.desc.d1.sspro.p2.Tsealed.center-d1.tau-target-T.ci95.1\endcsname{0}
\expandafter\def\csname pct@rev.desc.d1.sspro.p2.Tsealed.center-d1.tau-target-T.ci95.1\endcsname{24.0}
\expandafter\def\csname ppt@rev.desc.d1.sspro.p2.Tsealed.center-d1.tau-target-T.ci95.1\endcsname{$+$24.0}
\expandafter\def\csname val@rev.desc.d1.sspro.p2.Tsealed.center-d1.tau-target-T.mean\endcsname{0.156}
\expandafter\def\csname cnt@rev.desc.d1.sspro.p2.Tsealed.center-d1.tau-target-T.mean\endcsname{0}
\expandafter\def\csname pct@rev.desc.d1.sspro.p2.Tsealed.center-d1.tau-target-T.mean\endcsname{15.6}
\expandafter\def\csname ppt@rev.desc.d1.sspro.p2.Tsealed.center-d1.tau-target-T.mean\endcsname{$+$15.6}
\expandafter\def\csname val@rev.desc.d1.sspro.p2.Tsealed.center-d1.tau-target-T.n-apps\endcsname{12}
\expandafter\def\csname cnt@rev.desc.d1.sspro.p2.Tsealed.center-d1.tau-target-T.n-apps\endcsname{12}
\expandafter\def\csname pct@rev.desc.d1.sspro.p2.Tsealed.center-d1.tau-target-T.n-apps\endcsname{1200.0}
\expandafter\def\csname ppt@rev.desc.d1.sspro.p2.Tsealed.center-d1.tau-target-T.n-apps\endcsname{$+$1200.0}
\expandafter\def\csname val@rev.desc.d1.sspro.p2.Tsealed.center-d1.tau-target-T.tost1.ci90-inside\endcsname{0}
\expandafter\def\csname cnt@rev.desc.d1.sspro.p2.Tsealed.center-d1.tau-target-T.tost1.ci90-inside\endcsname{0}
\expandafter\def\csname pct@rev.desc.d1.sspro.p2.Tsealed.center-d1.tau-target-T.tost1.ci90-inside\endcsname{0.0}
\expandafter\def\csname ppt@rev.desc.d1.sspro.p2.Tsealed.center-d1.tau-target-T.tost1.ci90-inside\endcsname{0.0}
\expandafter\def\csname val@rev.desc.d1.sspro.p2.Tsealed.center-d1.tau-target-T.tost1.equivalent\endcsname{0}
\expandafter\def\csname cnt@rev.desc.d1.sspro.p2.Tsealed.center-d1.tau-target-T.tost1.equivalent\endcsname{0}
\expandafter\def\csname pct@rev.desc.d1.sspro.p2.Tsealed.center-d1.tau-target-T.tost1.equivalent\endcsname{0.0}
\expandafter\def\csname ppt@rev.desc.d1.sspro.p2.Tsealed.center-d1.tau-target-T.tost1.equivalent\endcsname{0.0}
\expandafter\def\csname val@rev.desc.d1.sspro.p2.Tsealed.center-d1.tau-target-T.tost1.p\endcsname{1.000}
\expandafter\def\csname cnt@rev.desc.d1.sspro.p2.Tsealed.center-d1.tau-target-T.tost1.p\endcsname{1}
\expandafter\def\csname pct@rev.desc.d1.sspro.p2.Tsealed.center-d1.tau-target-T.tost1.p\endcsname{100.0}
\expandafter\def\csname ppt@rev.desc.d1.sspro.p2.Tsealed.center-d1.tau-target-T.tost1.p\endcsname{$+$100.0}
\expandafter\def\csname val@rev.desc.d1.sspro.p2.Tsealed.center-d1.tau-target-T.tost5.ci90-inside\endcsname{0}
\expandafter\def\csname cnt@rev.desc.d1.sspro.p2.Tsealed.center-d1.tau-target-T.tost5.ci90-inside\endcsname{0}
\expandafter\def\csname pct@rev.desc.d1.sspro.p2.Tsealed.center-d1.tau-target-T.tost5.ci90-inside\endcsname{0.0}
\expandafter\def\csname ppt@rev.desc.d1.sspro.p2.Tsealed.center-d1.tau-target-T.tost5.ci90-inside\endcsname{0.0}
\expandafter\def\csname val@rev.desc.d1.sspro.p2.Tsealed.center-d1.tau-target-T.tost5.equivalent\endcsname{0}
\expandafter\def\csname cnt@rev.desc.d1.sspro.p2.Tsealed.center-d1.tau-target-T.tost5.equivalent\endcsname{0}
\expandafter\def\csname pct@rev.desc.d1.sspro.p2.Tsealed.center-d1.tau-target-T.tost5.equivalent\endcsname{0.0}
\expandafter\def\csname ppt@rev.desc.d1.sspro.p2.Tsealed.center-d1.tau-target-T.tost5.equivalent\endcsname{0.0}
\expandafter\def\csname val@rev.desc.d1.sspro.p2.Tsealed.center-d1.tau-target-T.tost5.p\endcsname{0.999}
\expandafter\def\csname cnt@rev.desc.d1.sspro.p2.Tsealed.center-d1.tau-target-T.tost5.p\endcsname{1}
\expandafter\def\csname pct@rev.desc.d1.sspro.p2.Tsealed.center-d1.tau-target-T.tost5.p\endcsname{99.9}
\expandafter\def\csname ppt@rev.desc.d1.sspro.p2.Tsealed.center-d1.tau-target-T.tost5.p\endcsname{$+$99.9}
\expandafter\def\csname val@rev.desc.d1.sspro.p2.Tsealed.context-d1.n-tasks\endcsname{96}
\expandafter\def\csname cnt@rev.desc.d1.sspro.p2.Tsealed.context-d1.n-tasks\endcsname{96}
\expandafter\def\csname pct@rev.desc.d1.sspro.p2.Tsealed.context-d1.n-tasks\endcsname{9600.0}
\expandafter\def\csname ppt@rev.desc.d1.sspro.p2.Tsealed.context-d1.n-tasks\endcsname{$+$9600.0}
\expandafter\def\csname val@rev.desc.d1.sspro.p2.Tsealed.context-d1.paper-tau-T\endcsname{0.125}
\expandafter\def\csname cnt@rev.desc.d1.sspro.p2.Tsealed.context-d1.paper-tau-T\endcsname{0}
\expandafter\def\csname pct@rev.desc.d1.sspro.p2.Tsealed.context-d1.paper-tau-T\endcsname{12.5}
\expandafter\def\csname ppt@rev.desc.d1.sspro.p2.Tsealed.context-d1.paper-tau-T\endcsname{$+$12.5}
\expandafter\def\csname val@rev.desc.d1.sspro.p2.Tsealed.context-d1.reproduces\endcsname{1}
\expandafter\def\csname cnt@rev.desc.d1.sspro.p2.Tsealed.context-d1.reproduces\endcsname{1}
\expandafter\def\csname pct@rev.desc.d1.sspro.p2.Tsealed.context-d1.reproduces\endcsname{100.0}
\expandafter\def\csname ppt@rev.desc.d1.sspro.p2.Tsealed.context-d1.reproduces\endcsname{$+$100.0}
\expandafter\def\csname val@rev.desc.d1.sspro.p2.Tsealed.context-d1.reproduces-ci\endcsname{1}
\expandafter\def\csname cnt@rev.desc.d1.sspro.p2.Tsealed.context-d1.reproduces-ci\endcsname{1}
\expandafter\def\csname pct@rev.desc.d1.sspro.p2.Tsealed.context-d1.reproduces-ci\endcsname{100.0}
\expandafter\def\csname ppt@rev.desc.d1.sspro.p2.Tsealed.context-d1.reproduces-ci\endcsname{$+$100.0}
\expandafter\def\csname val@rev.desc.d1.sspro.p2.Tsealed.context-d1.tau-target-T.ci90.0\endcsname{0.062}
\expandafter\def\csname cnt@rev.desc.d1.sspro.p2.Tsealed.context-d1.tau-target-T.ci90.0\endcsname{0}
\expandafter\def\csname pct@rev.desc.d1.sspro.p2.Tsealed.context-d1.tau-target-T.ci90.0\endcsname{6.2}
\expandafter\def\csname ppt@rev.desc.d1.sspro.p2.Tsealed.context-d1.tau-target-T.ci90.0\endcsname{$+$6.2}
\expandafter\def\csname val@rev.desc.d1.sspro.p2.Tsealed.context-d1.tau-target-T.ci90.1\endcsname{0.198}
\expandafter\def\csname cnt@rev.desc.d1.sspro.p2.Tsealed.context-d1.tau-target-T.ci90.1\endcsname{0}
\expandafter\def\csname pct@rev.desc.d1.sspro.p2.Tsealed.context-d1.tau-target-T.ci90.1\endcsname{19.8}
\expandafter\def\csname ppt@rev.desc.d1.sspro.p2.Tsealed.context-d1.tau-target-T.ci90.1\endcsname{$+$19.8}
\expandafter\def\csname val@rev.desc.d1.sspro.p2.Tsealed.context-d1.tau-target-T.ci95-excludes-0\endcsname{1}
\expandafter\def\csname cnt@rev.desc.d1.sspro.p2.Tsealed.context-d1.tau-target-T.ci95-excludes-0\endcsname{1}
\expandafter\def\csname pct@rev.desc.d1.sspro.p2.Tsealed.context-d1.tau-target-T.ci95-excludes-0\endcsname{100.0}
\expandafter\def\csname ppt@rev.desc.d1.sspro.p2.Tsealed.context-d1.tau-target-T.ci95-excludes-0\endcsname{$+$100.0}
\expandafter\def\csname val@rev.desc.d1.sspro.p2.Tsealed.context-d1.tau-target-T.ci95.0\endcsname{0.062}
\expandafter\def\csname cnt@rev.desc.d1.sspro.p2.Tsealed.context-d1.tau-target-T.ci95.0\endcsname{0}
\expandafter\def\csname pct@rev.desc.d1.sspro.p2.Tsealed.context-d1.tau-target-T.ci95.0\endcsname{6.2}
\expandafter\def\csname ppt@rev.desc.d1.sspro.p2.Tsealed.context-d1.tau-target-T.ci95.0\endcsname{$+$6.2}
\expandafter\def\csname val@rev.desc.d1.sspro.p2.Tsealed.context-d1.tau-target-T.ci95.1\endcsname{0.208}
\expandafter\def\csname cnt@rev.desc.d1.sspro.p2.Tsealed.context-d1.tau-target-T.ci95.1\endcsname{0}
\expandafter\def\csname pct@rev.desc.d1.sspro.p2.Tsealed.context-d1.tau-target-T.ci95.1\endcsname{20.8}
\expandafter\def\csname ppt@rev.desc.d1.sspro.p2.Tsealed.context-d1.tau-target-T.ci95.1\endcsname{$+$20.8}
\expandafter\def\csname val@rev.desc.d1.sspro.p2.Tsealed.context-d1.tau-target-T.mean\endcsname{0.125}
\expandafter\def\csname cnt@rev.desc.d1.sspro.p2.Tsealed.context-d1.tau-target-T.mean\endcsname{0}
\expandafter\def\csname pct@rev.desc.d1.sspro.p2.Tsealed.context-d1.tau-target-T.mean\endcsname{12.5}
\expandafter\def\csname ppt@rev.desc.d1.sspro.p2.Tsealed.context-d1.tau-target-T.mean\endcsname{$+$12.5}
\expandafter\def\csname val@rev.desc.d1.sspro.p2.Tsealed.context-d1.tau-target-T.n-apps\endcsname{12}
\expandafter\def\csname cnt@rev.desc.d1.sspro.p2.Tsealed.context-d1.tau-target-T.n-apps\endcsname{12}
\expandafter\def\csname pct@rev.desc.d1.sspro.p2.Tsealed.context-d1.tau-target-T.n-apps\endcsname{1200.0}
\expandafter\def\csname ppt@rev.desc.d1.sspro.p2.Tsealed.context-d1.tau-target-T.n-apps\endcsname{$+$1200.0}
\expandafter\def\csname val@rev.desc.d1.sspro.p2.Tsealed.context-d1.tau-target-T.tost1.ci90-inside\endcsname{0}
\expandafter\def\csname cnt@rev.desc.d1.sspro.p2.Tsealed.context-d1.tau-target-T.tost1.ci90-inside\endcsname{0}
\expandafter\def\csname pct@rev.desc.d1.sspro.p2.Tsealed.context-d1.tau-target-T.tost1.ci90-inside\endcsname{0.0}
\expandafter\def\csname ppt@rev.desc.d1.sspro.p2.Tsealed.context-d1.tau-target-T.tost1.ci90-inside\endcsname{0.0}
\expandafter\def\csname val@rev.desc.d1.sspro.p2.Tsealed.context-d1.tau-target-T.tost1.equivalent\endcsname{0}
\expandafter\def\csname cnt@rev.desc.d1.sspro.p2.Tsealed.context-d1.tau-target-T.tost1.equivalent\endcsname{0}
\expandafter\def\csname pct@rev.desc.d1.sspro.p2.Tsealed.context-d1.tau-target-T.tost1.equivalent\endcsname{0.0}
\expandafter\def\csname ppt@rev.desc.d1.sspro.p2.Tsealed.context-d1.tau-target-T.tost1.equivalent\endcsname{0.0}
\expandafter\def\csname val@rev.desc.d1.sspro.p2.Tsealed.context-d1.tau-target-T.tost1.p\endcsname{1.000}
\expandafter\def\csname cnt@rev.desc.d1.sspro.p2.Tsealed.context-d1.tau-target-T.tost1.p\endcsname{1}
\expandafter\def\csname pct@rev.desc.d1.sspro.p2.Tsealed.context-d1.tau-target-T.tost1.p\endcsname{100.0}
\expandafter\def\csname ppt@rev.desc.d1.sspro.p2.Tsealed.context-d1.tau-target-T.tost1.p\endcsname{$+$100.0}
\expandafter\def\csname val@rev.desc.d1.sspro.p2.Tsealed.context-d1.tau-target-T.tost5.ci90-inside\endcsname{0}
\expandafter\def\csname cnt@rev.desc.d1.sspro.p2.Tsealed.context-d1.tau-target-T.tost5.ci90-inside\endcsname{0}
\expandafter\def\csname pct@rev.desc.d1.sspro.p2.Tsealed.context-d1.tau-target-T.tost5.ci90-inside\endcsname{0.0}
\expandafter\def\csname ppt@rev.desc.d1.sspro.p2.Tsealed.context-d1.tau-target-T.tost5.ci90-inside\endcsname{0.0}
\expandafter\def\csname val@rev.desc.d1.sspro.p2.Tsealed.context-d1.tau-target-T.tost5.equivalent\endcsname{0}
\expandafter\def\csname cnt@rev.desc.d1.sspro.p2.Tsealed.context-d1.tau-target-T.tost5.equivalent\endcsname{0}
\expandafter\def\csname pct@rev.desc.d1.sspro.p2.Tsealed.context-d1.tau-target-T.tost5.equivalent\endcsname{0.0}
\expandafter\def\csname ppt@rev.desc.d1.sspro.p2.Tsealed.context-d1.tau-target-T.tost5.equivalent\endcsname{0.0}
\expandafter\def\csname val@rev.desc.d1.sspro.p2.Tsealed.context-d1.tau-target-T.tost5.p\endcsname{0.993}
\expandafter\def\csname cnt@rev.desc.d1.sspro.p2.Tsealed.context-d1.tau-target-T.tost5.p\endcsname{1}
\expandafter\def\csname pct@rev.desc.d1.sspro.p2.Tsealed.context-d1.tau-target-T.tost5.p\endcsname{99.3}
\expandafter\def\csname ppt@rev.desc.d1.sspro.p2.Tsealed.context-d1.tau-target-T.tost5.p\endcsname{$+$99.3}
\expandafter\def\csname val@rev.desc.d1.sspro.p2.Tsealed.dual-d2.n-tasks\endcsname{96}
\expandafter\def\csname cnt@rev.desc.d1.sspro.p2.Tsealed.dual-d2.n-tasks\endcsname{96}
\expandafter\def\csname pct@rev.desc.d1.sspro.p2.Tsealed.dual-d2.n-tasks\endcsname{9600.0}
\expandafter\def\csname ppt@rev.desc.d1.sspro.p2.Tsealed.dual-d2.n-tasks\endcsname{$+$9600.0}
\expandafter\def\csname val@rev.desc.d1.sspro.p2.Tsealed.dual-d2.paper-tau-T\endcsname{0.156}
\expandafter\def\csname cnt@rev.desc.d1.sspro.p2.Tsealed.dual-d2.paper-tau-T\endcsname{0}
\expandafter\def\csname pct@rev.desc.d1.sspro.p2.Tsealed.dual-d2.paper-tau-T\endcsname{15.6}
\expandafter\def\csname ppt@rev.desc.d1.sspro.p2.Tsealed.dual-d2.paper-tau-T\endcsname{$+$15.6}
\expandafter\def\csname val@rev.desc.d1.sspro.p2.Tsealed.dual-d2.reproduces\endcsname{1}
\expandafter\def\csname cnt@rev.desc.d1.sspro.p2.Tsealed.dual-d2.reproduces\endcsname{1}
\expandafter\def\csname pct@rev.desc.d1.sspro.p2.Tsealed.dual-d2.reproduces\endcsname{100.0}
\expandafter\def\csname ppt@rev.desc.d1.sspro.p2.Tsealed.dual-d2.reproduces\endcsname{$+$100.0}
\expandafter\def\csname val@rev.desc.d1.sspro.p2.Tsealed.dual-d2.reproduces-ci\endcsname{1}
\expandafter\def\csname cnt@rev.desc.d1.sspro.p2.Tsealed.dual-d2.reproduces-ci\endcsname{1}
\expandafter\def\csname pct@rev.desc.d1.sspro.p2.Tsealed.dual-d2.reproduces-ci\endcsname{100.0}
\expandafter\def\csname ppt@rev.desc.d1.sspro.p2.Tsealed.dual-d2.reproduces-ci\endcsname{$+$100.0}
\expandafter\def\csname val@rev.desc.d1.sspro.p2.Tsealed.dual-d2.tau-target-T.ci90.0\endcsname{0.094}
\expandafter\def\csname cnt@rev.desc.d1.sspro.p2.Tsealed.dual-d2.tau-target-T.ci90.0\endcsname{0}
\expandafter\def\csname pct@rev.desc.d1.sspro.p2.Tsealed.dual-d2.tau-target-T.ci90.0\endcsname{9.4}
\expandafter\def\csname ppt@rev.desc.d1.sspro.p2.Tsealed.dual-d2.tau-target-T.ci90.0\endcsname{$+$9.4}
\expandafter\def\csname val@rev.desc.d1.sspro.p2.Tsealed.dual-d2.tau-target-T.ci90.1\endcsname{0.219}
\expandafter\def\csname cnt@rev.desc.d1.sspro.p2.Tsealed.dual-d2.tau-target-T.ci90.1\endcsname{0}
\expandafter\def\csname pct@rev.desc.d1.sspro.p2.Tsealed.dual-d2.tau-target-T.ci90.1\endcsname{21.9}
\expandafter\def\csname ppt@rev.desc.d1.sspro.p2.Tsealed.dual-d2.tau-target-T.ci90.1\endcsname{$+$21.9}
\expandafter\def\csname val@rev.desc.d1.sspro.p2.Tsealed.dual-d2.tau-target-T.ci95-excludes-0\endcsname{1}
\expandafter\def\csname cnt@rev.desc.d1.sspro.p2.Tsealed.dual-d2.tau-target-T.ci95-excludes-0\endcsname{1}
\expandafter\def\csname pct@rev.desc.d1.sspro.p2.Tsealed.dual-d2.tau-target-T.ci95-excludes-0\endcsname{100.0}
\expandafter\def\csname ppt@rev.desc.d1.sspro.p2.Tsealed.dual-d2.tau-target-T.ci95-excludes-0\endcsname{$+$100.0}
\expandafter\def\csname val@rev.desc.d1.sspro.p2.Tsealed.dual-d2.tau-target-T.ci95.0\endcsname{0.083}
\expandafter\def\csname cnt@rev.desc.d1.sspro.p2.Tsealed.dual-d2.tau-target-T.ci95.0\endcsname{0}
\expandafter\def\csname pct@rev.desc.d1.sspro.p2.Tsealed.dual-d2.tau-target-T.ci95.0\endcsname{8.3}
\expandafter\def\csname ppt@rev.desc.d1.sspro.p2.Tsealed.dual-d2.tau-target-T.ci95.0\endcsname{$+$8.3}
\expandafter\def\csname val@rev.desc.d1.sspro.p2.Tsealed.dual-d2.tau-target-T.ci95.1\endcsname{0.229}
\expandafter\def\csname cnt@rev.desc.d1.sspro.p2.Tsealed.dual-d2.tau-target-T.ci95.1\endcsname{0}
\expandafter\def\csname pct@rev.desc.d1.sspro.p2.Tsealed.dual-d2.tau-target-T.ci95.1\endcsname{22.9}
\expandafter\def\csname ppt@rev.desc.d1.sspro.p2.Tsealed.dual-d2.tau-target-T.ci95.1\endcsname{$+$22.9}
\expandafter\def\csname val@rev.desc.d1.sspro.p2.Tsealed.dual-d2.tau-target-T.mean\endcsname{0.156}
\expandafter\def\csname cnt@rev.desc.d1.sspro.p2.Tsealed.dual-d2.tau-target-T.mean\endcsname{0}
\expandafter\def\csname pct@rev.desc.d1.sspro.p2.Tsealed.dual-d2.tau-target-T.mean\endcsname{15.6}
\expandafter\def\csname ppt@rev.desc.d1.sspro.p2.Tsealed.dual-d2.tau-target-T.mean\endcsname{$+$15.6}
\expandafter\def\csname val@rev.desc.d1.sspro.p2.Tsealed.dual-d2.tau-target-T.n-apps\endcsname{12}
\expandafter\def\csname cnt@rev.desc.d1.sspro.p2.Tsealed.dual-d2.tau-target-T.n-apps\endcsname{12}
\expandafter\def\csname pct@rev.desc.d1.sspro.p2.Tsealed.dual-d2.tau-target-T.n-apps\endcsname{1200.0}
\expandafter\def\csname ppt@rev.desc.d1.sspro.p2.Tsealed.dual-d2.tau-target-T.n-apps\endcsname{$+$1200.0}
\expandafter\def\csname val@rev.desc.d1.sspro.p2.Tsealed.dual-d2.tau-target-T.tost1.ci90-inside\endcsname{0}
\expandafter\def\csname cnt@rev.desc.d1.sspro.p2.Tsealed.dual-d2.tau-target-T.tost1.ci90-inside\endcsname{0}
\expandafter\def\csname pct@rev.desc.d1.sspro.p2.Tsealed.dual-d2.tau-target-T.tost1.ci90-inside\endcsname{0.0}
\expandafter\def\csname ppt@rev.desc.d1.sspro.p2.Tsealed.dual-d2.tau-target-T.tost1.ci90-inside\endcsname{0.0}
\expandafter\def\csname val@rev.desc.d1.sspro.p2.Tsealed.dual-d2.tau-target-T.tost1.equivalent\endcsname{0}
\expandafter\def\csname cnt@rev.desc.d1.sspro.p2.Tsealed.dual-d2.tau-target-T.tost1.equivalent\endcsname{0}
\expandafter\def\csname pct@rev.desc.d1.sspro.p2.Tsealed.dual-d2.tau-target-T.tost1.equivalent\endcsname{0.0}
\expandafter\def\csname ppt@rev.desc.d1.sspro.p2.Tsealed.dual-d2.tau-target-T.tost1.equivalent\endcsname{0.0}
\expandafter\def\csname val@rev.desc.d1.sspro.p2.Tsealed.dual-d2.tau-target-T.tost1.p\endcsname{1.000}
\expandafter\def\csname cnt@rev.desc.d1.sspro.p2.Tsealed.dual-d2.tau-target-T.tost1.p\endcsname{1}
\expandafter\def\csname pct@rev.desc.d1.sspro.p2.Tsealed.dual-d2.tau-target-T.tost1.p\endcsname{100.0}
\expandafter\def\csname ppt@rev.desc.d1.sspro.p2.Tsealed.dual-d2.tau-target-T.tost1.p\endcsname{$+$100.0}
\expandafter\def\csname val@rev.desc.d1.sspro.p2.Tsealed.dual-d2.tau-target-T.tost5.ci90-inside\endcsname{0}
\expandafter\def\csname cnt@rev.desc.d1.sspro.p2.Tsealed.dual-d2.tau-target-T.tost5.ci90-inside\endcsname{0}
\expandafter\def\csname pct@rev.desc.d1.sspro.p2.Tsealed.dual-d2.tau-target-T.tost5.ci90-inside\endcsname{0.0}
\expandafter\def\csname ppt@rev.desc.d1.sspro.p2.Tsealed.dual-d2.tau-target-T.tost5.ci90-inside\endcsname{0.0}
\expandafter\def\csname val@rev.desc.d1.sspro.p2.Tsealed.dual-d2.tau-target-T.tost5.equivalent\endcsname{0}
\expandafter\def\csname cnt@rev.desc.d1.sspro.p2.Tsealed.dual-d2.tau-target-T.tost5.equivalent\endcsname{0}
\expandafter\def\csname pct@rev.desc.d1.sspro.p2.Tsealed.dual-d2.tau-target-T.tost5.equivalent\endcsname{0.0}
\expandafter\def\csname ppt@rev.desc.d1.sspro.p2.Tsealed.dual-d2.tau-target-T.tost5.equivalent\endcsname{0.0}
\expandafter\def\csname val@rev.desc.d1.sspro.p2.Tsealed.dual-d2.tau-target-T.tost5.p\endcsname{1.000}
\expandafter\def\csname cnt@rev.desc.d1.sspro.p2.Tsealed.dual-d2.tau-target-T.tost5.p\endcsname{1}
\expandafter\def\csname pct@rev.desc.d1.sspro.p2.Tsealed.dual-d2.tau-target-T.tost5.p\endcsname{100.0}
\expandafter\def\csname ppt@rev.desc.d1.sspro.p2.Tsealed.dual-d2.tau-target-T.tost5.p\endcsname{$+$100.0}
\expandafter\def\csname val@rev.desc.d1.sspro.p2.Tsealed.grid2-d3.n-tasks\endcsname{96}
\expandafter\def\csname cnt@rev.desc.d1.sspro.p2.Tsealed.grid2-d3.n-tasks\endcsname{96}
\expandafter\def\csname pct@rev.desc.d1.sspro.p2.Tsealed.grid2-d3.n-tasks\endcsname{9600.0}
\expandafter\def\csname ppt@rev.desc.d1.sspro.p2.Tsealed.grid2-d3.n-tasks\endcsname{$+$9600.0}
\expandafter\def\csname val@rev.desc.d1.sspro.p2.Tsealed.grid2-d3.paper-tau-T\endcsname{0.146}
\expandafter\def\csname cnt@rev.desc.d1.sspro.p2.Tsealed.grid2-d3.paper-tau-T\endcsname{0}
\expandafter\def\csname pct@rev.desc.d1.sspro.p2.Tsealed.grid2-d3.paper-tau-T\endcsname{14.6}
\expandafter\def\csname ppt@rev.desc.d1.sspro.p2.Tsealed.grid2-d3.paper-tau-T\endcsname{$+$14.6}
\expandafter\def\csname val@rev.desc.d1.sspro.p2.Tsealed.grid2-d3.reproduces\endcsname{1}
\expandafter\def\csname cnt@rev.desc.d1.sspro.p2.Tsealed.grid2-d3.reproduces\endcsname{1}
\expandafter\def\csname pct@rev.desc.d1.sspro.p2.Tsealed.grid2-d3.reproduces\endcsname{100.0}
\expandafter\def\csname ppt@rev.desc.d1.sspro.p2.Tsealed.grid2-d3.reproduces\endcsname{$+$100.0}
\expandafter\def\csname val@rev.desc.d1.sspro.p2.Tsealed.grid2-d3.reproduces-ci\endcsname{1}
\expandafter\def\csname cnt@rev.desc.d1.sspro.p2.Tsealed.grid2-d3.reproduces-ci\endcsname{1}
\expandafter\def\csname pct@rev.desc.d1.sspro.p2.Tsealed.grid2-d3.reproduces-ci\endcsname{100.0}
\expandafter\def\csname ppt@rev.desc.d1.sspro.p2.Tsealed.grid2-d3.reproduces-ci\endcsname{$+$100.0}
\expandafter\def\csname val@rev.desc.d1.sspro.p2.Tsealed.grid2-d3.tau-target-T.ci90.0\endcsname{0.073}
\expandafter\def\csname cnt@rev.desc.d1.sspro.p2.Tsealed.grid2-d3.tau-target-T.ci90.0\endcsname{0}
\expandafter\def\csname pct@rev.desc.d1.sspro.p2.Tsealed.grid2-d3.tau-target-T.ci90.0\endcsname{7.3}
\expandafter\def\csname ppt@rev.desc.d1.sspro.p2.Tsealed.grid2-d3.tau-target-T.ci90.0\endcsname{$+$7.3}
\expandafter\def\csname val@rev.desc.d1.sspro.p2.Tsealed.grid2-d3.tau-target-T.ci90.1\endcsname{0.219}
\expandafter\def\csname cnt@rev.desc.d1.sspro.p2.Tsealed.grid2-d3.tau-target-T.ci90.1\endcsname{0}
\expandafter\def\csname pct@rev.desc.d1.sspro.p2.Tsealed.grid2-d3.tau-target-T.ci90.1\endcsname{21.9}
\expandafter\def\csname ppt@rev.desc.d1.sspro.p2.Tsealed.grid2-d3.tau-target-T.ci90.1\endcsname{$+$21.9}
\expandafter\def\csname val@rev.desc.d1.sspro.p2.Tsealed.grid2-d3.tau-target-T.ci95-excludes-0\endcsname{1}
\expandafter\def\csname cnt@rev.desc.d1.sspro.p2.Tsealed.grid2-d3.tau-target-T.ci95-excludes-0\endcsname{1}
\expandafter\def\csname pct@rev.desc.d1.sspro.p2.Tsealed.grid2-d3.tau-target-T.ci95-excludes-0\endcsname{100.0}
\expandafter\def\csname ppt@rev.desc.d1.sspro.p2.Tsealed.grid2-d3.tau-target-T.ci95-excludes-0\endcsname{$+$100.0}
\expandafter\def\csname val@rev.desc.d1.sspro.p2.Tsealed.grid2-d3.tau-target-T.ci95.0\endcsname{0.062}
\expandafter\def\csname cnt@rev.desc.d1.sspro.p2.Tsealed.grid2-d3.tau-target-T.ci95.0\endcsname{0}
\expandafter\def\csname pct@rev.desc.d1.sspro.p2.Tsealed.grid2-d3.tau-target-T.ci95.0\endcsname{6.2}
\expandafter\def\csname ppt@rev.desc.d1.sspro.p2.Tsealed.grid2-d3.tau-target-T.ci95.0\endcsname{$+$6.2}
\expandafter\def\csname val@rev.desc.d1.sspro.p2.Tsealed.grid2-d3.tau-target-T.ci95.1\endcsname{0.240}
\expandafter\def\csname cnt@rev.desc.d1.sspro.p2.Tsealed.grid2-d3.tau-target-T.ci95.1\endcsname{0}
\expandafter\def\csname pct@rev.desc.d1.sspro.p2.Tsealed.grid2-d3.tau-target-T.ci95.1\endcsname{24.0}
\expandafter\def\csname ppt@rev.desc.d1.sspro.p2.Tsealed.grid2-d3.tau-target-T.ci95.1\endcsname{$+$24.0}
\expandafter\def\csname val@rev.desc.d1.sspro.p2.Tsealed.grid2-d3.tau-target-T.mean\endcsname{0.146}
\expandafter\def\csname cnt@rev.desc.d1.sspro.p2.Tsealed.grid2-d3.tau-target-T.mean\endcsname{0}
\expandafter\def\csname pct@rev.desc.d1.sspro.p2.Tsealed.grid2-d3.tau-target-T.mean\endcsname{14.6}
\expandafter\def\csname ppt@rev.desc.d1.sspro.p2.Tsealed.grid2-d3.tau-target-T.mean\endcsname{$+$14.6}
\expandafter\def\csname val@rev.desc.d1.sspro.p2.Tsealed.grid2-d3.tau-target-T.n-apps\endcsname{12}
\expandafter\def\csname cnt@rev.desc.d1.sspro.p2.Tsealed.grid2-d3.tau-target-T.n-apps\endcsname{12}
\expandafter\def\csname pct@rev.desc.d1.sspro.p2.Tsealed.grid2-d3.tau-target-T.n-apps\endcsname{1200.0}
\expandafter\def\csname ppt@rev.desc.d1.sspro.p2.Tsealed.grid2-d3.tau-target-T.n-apps\endcsname{$+$1200.0}
\expandafter\def\csname val@rev.desc.d1.sspro.p2.Tsealed.grid2-d3.tau-target-T.tost1.ci90-inside\endcsname{0}
\expandafter\def\csname cnt@rev.desc.d1.sspro.p2.Tsealed.grid2-d3.tau-target-T.tost1.ci90-inside\endcsname{0}
\expandafter\def\csname pct@rev.desc.d1.sspro.p2.Tsealed.grid2-d3.tau-target-T.tost1.ci90-inside\endcsname{0.0}
\expandafter\def\csname ppt@rev.desc.d1.sspro.p2.Tsealed.grid2-d3.tau-target-T.tost1.ci90-inside\endcsname{0.0}
\expandafter\def\csname val@rev.desc.d1.sspro.p2.Tsealed.grid2-d3.tau-target-T.tost1.equivalent\endcsname{0}
\expandafter\def\csname cnt@rev.desc.d1.sspro.p2.Tsealed.grid2-d3.tau-target-T.tost1.equivalent\endcsname{0}
\expandafter\def\csname pct@rev.desc.d1.sspro.p2.Tsealed.grid2-d3.tau-target-T.tost1.equivalent\endcsname{0.0}
\expandafter\def\csname ppt@rev.desc.d1.sspro.p2.Tsealed.grid2-d3.tau-target-T.tost1.equivalent\endcsname{0.0}
\expandafter\def\csname val@rev.desc.d1.sspro.p2.Tsealed.grid2-d3.tau-target-T.tost1.p\endcsname{1.000}
\expandafter\def\csname cnt@rev.desc.d1.sspro.p2.Tsealed.grid2-d3.tau-target-T.tost1.p\endcsname{1}
\expandafter\def\csname pct@rev.desc.d1.sspro.p2.Tsealed.grid2-d3.tau-target-T.tost1.p\endcsname{100.0}
\expandafter\def\csname ppt@rev.desc.d1.sspro.p2.Tsealed.grid2-d3.tau-target-T.tost1.p\endcsname{$+$100.0}
\expandafter\def\csname val@rev.desc.d1.sspro.p2.Tsealed.grid2-d3.tau-target-T.tost5.ci90-inside\endcsname{0}
\expandafter\def\csname cnt@rev.desc.d1.sspro.p2.Tsealed.grid2-d3.tau-target-T.tost5.ci90-inside\endcsname{0}
\expandafter\def\csname pct@rev.desc.d1.sspro.p2.Tsealed.grid2-d3.tau-target-T.tost5.ci90-inside\endcsname{0.0}
\expandafter\def\csname ppt@rev.desc.d1.sspro.p2.Tsealed.grid2-d3.tau-target-T.tost5.ci90-inside\endcsname{0.0}
\expandafter\def\csname val@rev.desc.d1.sspro.p2.Tsealed.grid2-d3.tau-target-T.tost5.equivalent\endcsname{0}
\expandafter\def\csname cnt@rev.desc.d1.sspro.p2.Tsealed.grid2-d3.tau-target-T.tost5.equivalent\endcsname{0}
\expandafter\def\csname pct@rev.desc.d1.sspro.p2.Tsealed.grid2-d3.tau-target-T.tost5.equivalent\endcsname{0.0}
\expandafter\def\csname ppt@rev.desc.d1.sspro.p2.Tsealed.grid2-d3.tau-target-T.tost5.equivalent\endcsname{0.0}
\expandafter\def\csname val@rev.desc.d1.sspro.p2.Tsealed.grid2-d3.tau-target-T.tost5.p\endcsname{0.992}
\expandafter\def\csname cnt@rev.desc.d1.sspro.p2.Tsealed.grid2-d3.tau-target-T.tost5.p\endcsname{1}
\expandafter\def\csname pct@rev.desc.d1.sspro.p2.Tsealed.grid2-d3.tau-target-T.tost5.p\endcsname{99.2}
\expandafter\def\csname ppt@rev.desc.d1.sspro.p2.Tsealed.grid2-d3.tau-target-T.tost5.p\endcsname{$+$99.2}
\expandafter\def\csname val@rev.desc.d1.sspro.p2.Tsealed.zoomclick-d3.n-tasks\endcsname{96}
\expandafter\def\csname cnt@rev.desc.d1.sspro.p2.Tsealed.zoomclick-d3.n-tasks\endcsname{96}
\expandafter\def\csname pct@rev.desc.d1.sspro.p2.Tsealed.zoomclick-d3.n-tasks\endcsname{9600.0}
\expandafter\def\csname ppt@rev.desc.d1.sspro.p2.Tsealed.zoomclick-d3.n-tasks\endcsname{$+$9600.0}
\expandafter\def\csname val@rev.desc.d1.sspro.p2.Tsealed.zoomclick-d3.paper-tau-T\endcsname{0.167}
\expandafter\def\csname cnt@rev.desc.d1.sspro.p2.Tsealed.zoomclick-d3.paper-tau-T\endcsname{0}
\expandafter\def\csname pct@rev.desc.d1.sspro.p2.Tsealed.zoomclick-d3.paper-tau-T\endcsname{16.7}
\expandafter\def\csname ppt@rev.desc.d1.sspro.p2.Tsealed.zoomclick-d3.paper-tau-T\endcsname{$+$16.7}
\expandafter\def\csname val@rev.desc.d1.sspro.p2.Tsealed.zoomclick-d3.reproduces\endcsname{1}
\expandafter\def\csname cnt@rev.desc.d1.sspro.p2.Tsealed.zoomclick-d3.reproduces\endcsname{1}
\expandafter\def\csname pct@rev.desc.d1.sspro.p2.Tsealed.zoomclick-d3.reproduces\endcsname{100.0}
\expandafter\def\csname ppt@rev.desc.d1.sspro.p2.Tsealed.zoomclick-d3.reproduces\endcsname{$+$100.0}
\expandafter\def\csname val@rev.desc.d1.sspro.p2.Tsealed.zoomclick-d3.reproduces-ci\endcsname{1}
\expandafter\def\csname cnt@rev.desc.d1.sspro.p2.Tsealed.zoomclick-d3.reproduces-ci\endcsname{1}
\expandafter\def\csname pct@rev.desc.d1.sspro.p2.Tsealed.zoomclick-d3.reproduces-ci\endcsname{100.0}
\expandafter\def\csname ppt@rev.desc.d1.sspro.p2.Tsealed.zoomclick-d3.reproduces-ci\endcsname{$+$100.0}
\expandafter\def\csname val@rev.desc.d1.sspro.p2.Tsealed.zoomclick-d3.tau-target-T.ci90.0\endcsname{0.115}
\expandafter\def\csname cnt@rev.desc.d1.sspro.p2.Tsealed.zoomclick-d3.tau-target-T.ci90.0\endcsname{0}
\expandafter\def\csname pct@rev.desc.d1.sspro.p2.Tsealed.zoomclick-d3.tau-target-T.ci90.0\endcsname{11.5}
\expandafter\def\csname ppt@rev.desc.d1.sspro.p2.Tsealed.zoomclick-d3.tau-target-T.ci90.0\endcsname{$+$11.5}
\expandafter\def\csname val@rev.desc.d1.sspro.p2.Tsealed.zoomclick-d3.tau-target-T.ci90.1\endcsname{0.229}
\expandafter\def\csname cnt@rev.desc.d1.sspro.p2.Tsealed.zoomclick-d3.tau-target-T.ci90.1\endcsname{0}
\expandafter\def\csname pct@rev.desc.d1.sspro.p2.Tsealed.zoomclick-d3.tau-target-T.ci90.1\endcsname{22.9}
\expandafter\def\csname ppt@rev.desc.d1.sspro.p2.Tsealed.zoomclick-d3.tau-target-T.ci90.1\endcsname{$+$22.9}
\expandafter\def\csname val@rev.desc.d1.sspro.p2.Tsealed.zoomclick-d3.tau-target-T.ci95-excludes-0\endcsname{1}
\expandafter\def\csname cnt@rev.desc.d1.sspro.p2.Tsealed.zoomclick-d3.tau-target-T.ci95-excludes-0\endcsname{1}
\expandafter\def\csname pct@rev.desc.d1.sspro.p2.Tsealed.zoomclick-d3.tau-target-T.ci95-excludes-0\endcsname{100.0}
\expandafter\def\csname ppt@rev.desc.d1.sspro.p2.Tsealed.zoomclick-d3.tau-target-T.ci95-excludes-0\endcsname{$+$100.0}
\expandafter\def\csname val@rev.desc.d1.sspro.p2.Tsealed.zoomclick-d3.tau-target-T.ci95.0\endcsname{0.104}
\expandafter\def\csname cnt@rev.desc.d1.sspro.p2.Tsealed.zoomclick-d3.tau-target-T.ci95.0\endcsname{0}
\expandafter\def\csname pct@rev.desc.d1.sspro.p2.Tsealed.zoomclick-d3.tau-target-T.ci95.0\endcsname{10.4}
\expandafter\def\csname ppt@rev.desc.d1.sspro.p2.Tsealed.zoomclick-d3.tau-target-T.ci95.0\endcsname{$+$10.4}
\expandafter\def\csname val@rev.desc.d1.sspro.p2.Tsealed.zoomclick-d3.tau-target-T.ci95.1\endcsname{0.240}
\expandafter\def\csname cnt@rev.desc.d1.sspro.p2.Tsealed.zoomclick-d3.tau-target-T.ci95.1\endcsname{0}
\expandafter\def\csname pct@rev.desc.d1.sspro.p2.Tsealed.zoomclick-d3.tau-target-T.ci95.1\endcsname{24.0}
\expandafter\def\csname ppt@rev.desc.d1.sspro.p2.Tsealed.zoomclick-d3.tau-target-T.ci95.1\endcsname{$+$24.0}
\expandafter\def\csname val@rev.desc.d1.sspro.p2.Tsealed.zoomclick-d3.tau-target-T.mean\endcsname{0.167}
\expandafter\def\csname cnt@rev.desc.d1.sspro.p2.Tsealed.zoomclick-d3.tau-target-T.mean\endcsname{0}
\expandafter\def\csname pct@rev.desc.d1.sspro.p2.Tsealed.zoomclick-d3.tau-target-T.mean\endcsname{16.7}
\expandafter\def\csname ppt@rev.desc.d1.sspro.p2.Tsealed.zoomclick-d3.tau-target-T.mean\endcsname{$+$16.7}
\expandafter\def\csname val@rev.desc.d1.sspro.p2.Tsealed.zoomclick-d3.tau-target-T.n-apps\endcsname{12}
\expandafter\def\csname cnt@rev.desc.d1.sspro.p2.Tsealed.zoomclick-d3.tau-target-T.n-apps\endcsname{12}
\expandafter\def\csname pct@rev.desc.d1.sspro.p2.Tsealed.zoomclick-d3.tau-target-T.n-apps\endcsname{1200.0}
\expandafter\def\csname ppt@rev.desc.d1.sspro.p2.Tsealed.zoomclick-d3.tau-target-T.n-apps\endcsname{$+$1200.0}
\expandafter\def\csname val@rev.desc.d1.sspro.p2.Tsealed.zoomclick-d3.tau-target-T.tost1.ci90-inside\endcsname{0}
\expandafter\def\csname cnt@rev.desc.d1.sspro.p2.Tsealed.zoomclick-d3.tau-target-T.tost1.ci90-inside\endcsname{0}
\expandafter\def\csname pct@rev.desc.d1.sspro.p2.Tsealed.zoomclick-d3.tau-target-T.tost1.ci90-inside\endcsname{0.0}
\expandafter\def\csname ppt@rev.desc.d1.sspro.p2.Tsealed.zoomclick-d3.tau-target-T.tost1.ci90-inside\endcsname{0.0}
\expandafter\def\csname val@rev.desc.d1.sspro.p2.Tsealed.zoomclick-d3.tau-target-T.tost1.equivalent\endcsname{0}
\expandafter\def\csname cnt@rev.desc.d1.sspro.p2.Tsealed.zoomclick-d3.tau-target-T.tost1.equivalent\endcsname{0}
\expandafter\def\csname pct@rev.desc.d1.sspro.p2.Tsealed.zoomclick-d3.tau-target-T.tost1.equivalent\endcsname{0.0}
\expandafter\def\csname ppt@rev.desc.d1.sspro.p2.Tsealed.zoomclick-d3.tau-target-T.tost1.equivalent\endcsname{0.0}
\expandafter\def\csname val@rev.desc.d1.sspro.p2.Tsealed.zoomclick-d3.tau-target-T.tost1.p\endcsname{1.000}
\expandafter\def\csname cnt@rev.desc.d1.sspro.p2.Tsealed.zoomclick-d3.tau-target-T.tost1.p\endcsname{1}
\expandafter\def\csname pct@rev.desc.d1.sspro.p2.Tsealed.zoomclick-d3.tau-target-T.tost1.p\endcsname{100.0}
\expandafter\def\csname ppt@rev.desc.d1.sspro.p2.Tsealed.zoomclick-d3.tau-target-T.tost1.p\endcsname{$+$100.0}
\expandafter\def\csname val@rev.desc.d1.sspro.p2.Tsealed.zoomclick-d3.tau-target-T.tost5.ci90-inside\endcsname{0}
\expandafter\def\csname cnt@rev.desc.d1.sspro.p2.Tsealed.zoomclick-d3.tau-target-T.tost5.ci90-inside\endcsname{0}
\expandafter\def\csname pct@rev.desc.d1.sspro.p2.Tsealed.zoomclick-d3.tau-target-T.tost5.ci90-inside\endcsname{0.0}
\expandafter\def\csname ppt@rev.desc.d1.sspro.p2.Tsealed.zoomclick-d3.tau-target-T.tost5.ci90-inside\endcsname{0.0}
\expandafter\def\csname val@rev.desc.d1.sspro.p2.Tsealed.zoomclick-d3.tau-target-T.tost5.equivalent\endcsname{0}
\expandafter\def\csname cnt@rev.desc.d1.sspro.p2.Tsealed.zoomclick-d3.tau-target-T.tost5.equivalent\endcsname{0}
\expandafter\def\csname pct@rev.desc.d1.sspro.p2.Tsealed.zoomclick-d3.tau-target-T.tost5.equivalent\endcsname{0.0}
\expandafter\def\csname ppt@rev.desc.d1.sspro.p2.Tsealed.zoomclick-d3.tau-target-T.tost5.equivalent\endcsname{0.0}
\expandafter\def\csname val@rev.desc.d1.sspro.p2.Tsealed.zoomclick-d3.tau-target-T.tost5.p\endcsname{1.000}
\expandafter\def\csname cnt@rev.desc.d1.sspro.p2.Tsealed.zoomclick-d3.tau-target-T.tost5.p\endcsname{1}
\expandafter\def\csname pct@rev.desc.d1.sspro.p2.Tsealed.zoomclick-d3.tau-target-T.tost5.p\endcsname{100.0}
\expandafter\def\csname ppt@rev.desc.d1.sspro.p2.Tsealed.zoomclick-d3.tau-target-T.tost5.p\endcsname{$+$100.0}
\expandafter\def\csname val@rev.desc.d1.uiv.p1.C.center-d1.delta-tau.ci90.0\endcsname{-0.021}
\expandafter\def\csname cnt@rev.desc.d1.uiv.p1.C.center-d1.delta-tau.ci90.0\endcsname{0}
\expandafter\def\csname pct@rev.desc.d1.uiv.p1.C.center-d1.delta-tau.ci90.0\endcsname{$-$2.1}
\expandafter\def\csname ppt@rev.desc.d1.uiv.p1.C.center-d1.delta-tau.ci90.0\endcsname{$-$2.1}
\expandafter\def\csname val@rev.desc.d1.uiv.p1.C.center-d1.delta-tau.ci90.1\endcsname{0.177}
\expandafter\def\csname cnt@rev.desc.d1.uiv.p1.C.center-d1.delta-tau.ci90.1\endcsname{0}
\expandafter\def\csname pct@rev.desc.d1.uiv.p1.C.center-d1.delta-tau.ci90.1\endcsname{17.7}
\expandafter\def\csname ppt@rev.desc.d1.uiv.p1.C.center-d1.delta-tau.ci90.1\endcsname{$+$17.7}
\expandafter\def\csname val@rev.desc.d1.uiv.p1.C.center-d1.delta-tau.ci95-excludes-0\endcsname{0}
\expandafter\def\csname cnt@rev.desc.d1.uiv.p1.C.center-d1.delta-tau.ci95-excludes-0\endcsname{0}
\expandafter\def\csname pct@rev.desc.d1.uiv.p1.C.center-d1.delta-tau.ci95-excludes-0\endcsname{0.0}
\expandafter\def\csname ppt@rev.desc.d1.uiv.p1.C.center-d1.delta-tau.ci95-excludes-0\endcsname{0.0}
\expandafter\def\csname val@rev.desc.d1.uiv.p1.C.center-d1.delta-tau.ci95.0\endcsname{-0.042}
\expandafter\def\csname cnt@rev.desc.d1.uiv.p1.C.center-d1.delta-tau.ci95.0\endcsname{0}
\expandafter\def\csname pct@rev.desc.d1.uiv.p1.C.center-d1.delta-tau.ci95.0\endcsname{$-$4.2}
\expandafter\def\csname ppt@rev.desc.d1.uiv.p1.C.center-d1.delta-tau.ci95.0\endcsname{$-$4.2}
\expandafter\def\csname val@rev.desc.d1.uiv.p1.C.center-d1.delta-tau.ci95.1\endcsname{0.198}
\expandafter\def\csname cnt@rev.desc.d1.uiv.p1.C.center-d1.delta-tau.ci95.1\endcsname{0}
\expandafter\def\csname pct@rev.desc.d1.uiv.p1.C.center-d1.delta-tau.ci95.1\endcsname{19.8}
\expandafter\def\csname ppt@rev.desc.d1.uiv.p1.C.center-d1.delta-tau.ci95.1\endcsname{$+$19.8}
\expandafter\def\csname val@rev.desc.d1.uiv.p1.C.center-d1.delta-tau.mean\endcsname{0.083}
\expandafter\def\csname cnt@rev.desc.d1.uiv.p1.C.center-d1.delta-tau.mean\endcsname{0}
\expandafter\def\csname pct@rev.desc.d1.uiv.p1.C.center-d1.delta-tau.mean\endcsname{8.3}
\expandafter\def\csname ppt@rev.desc.d1.uiv.p1.C.center-d1.delta-tau.mean\endcsname{$+$8.3}
\expandafter\def\csname val@rev.desc.d1.uiv.p1.C.center-d1.delta-tau.n-apps\endcsname{12}
\expandafter\def\csname cnt@rev.desc.d1.uiv.p1.C.center-d1.delta-tau.n-apps\endcsname{12}
\expandafter\def\csname pct@rev.desc.d1.uiv.p1.C.center-d1.delta-tau.n-apps\endcsname{1200.0}
\expandafter\def\csname ppt@rev.desc.d1.uiv.p1.C.center-d1.delta-tau.n-apps\endcsname{$+$1200.0}
\expandafter\def\csname val@rev.desc.d1.uiv.p1.C.center-d1.delta-tau.tost1.ci90-inside\endcsname{0}
\expandafter\def\csname cnt@rev.desc.d1.uiv.p1.C.center-d1.delta-tau.tost1.ci90-inside\endcsname{0}
\expandafter\def\csname pct@rev.desc.d1.uiv.p1.C.center-d1.delta-tau.tost1.ci90-inside\endcsname{0.0}
\expandafter\def\csname ppt@rev.desc.d1.uiv.p1.C.center-d1.delta-tau.tost1.ci90-inside\endcsname{0.0}
\expandafter\def\csname val@rev.desc.d1.uiv.p1.C.center-d1.delta-tau.tost1.equivalent\endcsname{0}
\expandafter\def\csname cnt@rev.desc.d1.uiv.p1.C.center-d1.delta-tau.tost1.equivalent\endcsname{0}
\expandafter\def\csname pct@rev.desc.d1.uiv.p1.C.center-d1.delta-tau.tost1.equivalent\endcsname{0.0}
\expandafter\def\csname ppt@rev.desc.d1.uiv.p1.C.center-d1.delta-tau.tost1.equivalent\endcsname{0.0}
\expandafter\def\csname val@rev.desc.d1.uiv.p1.C.center-d1.delta-tau.tost1.p\endcsname{0.897}
\expandafter\def\csname cnt@rev.desc.d1.uiv.p1.C.center-d1.delta-tau.tost1.p\endcsname{1}
\expandafter\def\csname pct@rev.desc.d1.uiv.p1.C.center-d1.delta-tau.tost1.p\endcsname{89.7}
\expandafter\def\csname ppt@rev.desc.d1.uiv.p1.C.center-d1.delta-tau.tost1.p\endcsname{$+$89.7}
\expandafter\def\csname val@rev.desc.d1.uiv.p1.C.center-d1.delta-tau.tost5.ci90-inside\endcsname{0}
\expandafter\def\csname cnt@rev.desc.d1.uiv.p1.C.center-d1.delta-tau.tost5.ci90-inside\endcsname{0}
\expandafter\def\csname pct@rev.desc.d1.uiv.p1.C.center-d1.delta-tau.tost5.ci90-inside\endcsname{0.0}
\expandafter\def\csname ppt@rev.desc.d1.uiv.p1.C.center-d1.delta-tau.tost5.ci90-inside\endcsname{0.0}
\expandafter\def\csname val@rev.desc.d1.uiv.p1.C.center-d1.delta-tau.tost5.equivalent\endcsname{0}
\expandafter\def\csname cnt@rev.desc.d1.uiv.p1.C.center-d1.delta-tau.tost5.equivalent\endcsname{0}
\expandafter\def\csname pct@rev.desc.d1.uiv.p1.C.center-d1.delta-tau.tost5.equivalent\endcsname{0.0}
\expandafter\def\csname ppt@rev.desc.d1.uiv.p1.C.center-d1.delta-tau.tost5.equivalent\endcsname{0.0}
\expandafter\def\csname val@rev.desc.d1.uiv.p1.C.center-d1.delta-tau.tost5.p\endcsname{0.733}
\expandafter\def\csname cnt@rev.desc.d1.uiv.p1.C.center-d1.delta-tau.tost5.p\endcsname{1}
\expandafter\def\csname pct@rev.desc.d1.uiv.p1.C.center-d1.delta-tau.tost5.p\endcsname{73.3}
\expandafter\def\csname ppt@rev.desc.d1.uiv.p1.C.center-d1.delta-tau.tost5.p\endcsname{$+$73.3}
\expandafter\def\csname val@rev.desc.d1.uiv.p1.C.center-d1.material-flip\endcsname{0}
\expandafter\def\csname cnt@rev.desc.d1.uiv.p1.C.center-d1.material-flip\endcsname{0}
\expandafter\def\csname pct@rev.desc.d1.uiv.p1.C.center-d1.material-flip\endcsname{0.0}
\expandafter\def\csname ppt@rev.desc.d1.uiv.p1.C.center-d1.material-flip\endcsname{0.0}
\expandafter\def\csname val@rev.desc.d1.uiv.p1.C.center-d1.n-apps\endcsname{12}
\expandafter\def\csname cnt@rev.desc.d1.uiv.p1.C.center-d1.n-apps\endcsname{12}
\expandafter\def\csname pct@rev.desc.d1.uiv.p1.C.center-d1.n-apps\endcsname{1200.0}
\expandafter\def\csname ppt@rev.desc.d1.uiv.p1.C.center-d1.n-apps\endcsname{$+$1200.0}
\expandafter\def\csname val@rev.desc.d1.uiv.p1.C.center-d1.n-tasks\endcsname{96}
\expandafter\def\csname cnt@rev.desc.d1.uiv.p1.C.center-d1.n-tasks\endcsname{96}
\expandafter\def\csname pct@rev.desc.d1.uiv.p1.C.center-d1.n-tasks\endcsname{9600.0}
\expandafter\def\csname ppt@rev.desc.d1.uiv.p1.C.center-d1.n-tasks\endcsname{$+$9600.0}
\expandafter\def\csname val@rev.desc.d1.uiv.p1.C.center-d1.sign-differs\endcsname{1}
\expandafter\def\csname cnt@rev.desc.d1.uiv.p1.C.center-d1.sign-differs\endcsname{1}
\expandafter\def\csname pct@rev.desc.d1.uiv.p1.C.center-d1.sign-differs\endcsname{100.0}
\expandafter\def\csname ppt@rev.desc.d1.uiv.p1.C.center-d1.sign-differs\endcsname{$+$100.0}
\expandafter\def\csname val@rev.desc.d1.uiv.p1.C.center-d1.tau-source.ci90.0\endcsname{-0.135}
\expandafter\def\csname cnt@rev.desc.d1.uiv.p1.C.center-d1.tau-source.ci90.0\endcsname{0}
\expandafter\def\csname pct@rev.desc.d1.uiv.p1.C.center-d1.tau-source.ci90.0\endcsname{$-$13.5}
\expandafter\def\csname ppt@rev.desc.d1.uiv.p1.C.center-d1.tau-source.ci90.0\endcsname{$-$13.5}
\expandafter\def\csname val@rev.desc.d1.uiv.p1.C.center-d1.tau-source.ci90.1\endcsname{-0.010}
\expandafter\def\csname cnt@rev.desc.d1.uiv.p1.C.center-d1.tau-source.ci90.1\endcsname{0}
\expandafter\def\csname pct@rev.desc.d1.uiv.p1.C.center-d1.tau-source.ci90.1\endcsname{$-$1.0}
\expandafter\def\csname ppt@rev.desc.d1.uiv.p1.C.center-d1.tau-source.ci90.1\endcsname{$-$1.0}
\expandafter\def\csname val@rev.desc.d1.uiv.p1.C.center-d1.tau-source.ci95-excludes-0\endcsname{0}
\expandafter\def\csname cnt@rev.desc.d1.uiv.p1.C.center-d1.tau-source.ci95-excludes-0\endcsname{0}
\expandafter\def\csname pct@rev.desc.d1.uiv.p1.C.center-d1.tau-source.ci95-excludes-0\endcsname{0.0}
\expandafter\def\csname ppt@rev.desc.d1.uiv.p1.C.center-d1.tau-source.ci95-excludes-0\endcsname{0.0}
\expandafter\def\csname val@rev.desc.d1.uiv.p1.C.center-d1.tau-source.ci95.0\endcsname{-0.146}
\expandafter\def\csname cnt@rev.desc.d1.uiv.p1.C.center-d1.tau-source.ci95.0\endcsname{0}
\expandafter\def\csname pct@rev.desc.d1.uiv.p1.C.center-d1.tau-source.ci95.0\endcsname{$-$14.6}
\expandafter\def\csname ppt@rev.desc.d1.uiv.p1.C.center-d1.tau-source.ci95.0\endcsname{$-$14.6}
\expandafter\def\csname val@rev.desc.d1.uiv.p1.C.center-d1.tau-source.ci95.1\endcsname{0.000}
\expandafter\def\csname cnt@rev.desc.d1.uiv.p1.C.center-d1.tau-source.ci95.1\endcsname{0}
\expandafter\def\csname pct@rev.desc.d1.uiv.p1.C.center-d1.tau-source.ci95.1\endcsname{0.0}
\expandafter\def\csname ppt@rev.desc.d1.uiv.p1.C.center-d1.tau-source.ci95.1\endcsname{0.0}
\expandafter\def\csname val@rev.desc.d1.uiv.p1.C.center-d1.tau-source.mean\endcsname{-0.073}
\expandafter\def\csname cnt@rev.desc.d1.uiv.p1.C.center-d1.tau-source.mean\endcsname{0}
\expandafter\def\csname pct@rev.desc.d1.uiv.p1.C.center-d1.tau-source.mean\endcsname{$-$7.3}
\expandafter\def\csname ppt@rev.desc.d1.uiv.p1.C.center-d1.tau-source.mean\endcsname{$-$7.3}
\expandafter\def\csname val@rev.desc.d1.uiv.p1.C.center-d1.tau-source.n-apps\endcsname{12}
\expandafter\def\csname cnt@rev.desc.d1.uiv.p1.C.center-d1.tau-source.n-apps\endcsname{12}
\expandafter\def\csname pct@rev.desc.d1.uiv.p1.C.center-d1.tau-source.n-apps\endcsname{1200.0}
\expandafter\def\csname ppt@rev.desc.d1.uiv.p1.C.center-d1.tau-source.n-apps\endcsname{$+$1200.0}
\expandafter\def\csname val@rev.desc.d1.uiv.p1.C.center-d1.tau-source.tost1.ci90-inside\endcsname{0}
\expandafter\def\csname cnt@rev.desc.d1.uiv.p1.C.center-d1.tau-source.tost1.ci90-inside\endcsname{0}
\expandafter\def\csname pct@rev.desc.d1.uiv.p1.C.center-d1.tau-source.tost1.ci90-inside\endcsname{0.0}
\expandafter\def\csname ppt@rev.desc.d1.uiv.p1.C.center-d1.tau-source.tost1.ci90-inside\endcsname{0.0}
\expandafter\def\csname val@rev.desc.d1.uiv.p1.C.center-d1.tau-source.tost1.equivalent\endcsname{0}
\expandafter\def\csname cnt@rev.desc.d1.uiv.p1.C.center-d1.tau-source.tost1.equivalent\endcsname{0}
\expandafter\def\csname pct@rev.desc.d1.uiv.p1.C.center-d1.tau-source.tost1.equivalent\endcsname{0.0}
\expandafter\def\csname ppt@rev.desc.d1.uiv.p1.C.center-d1.tau-source.tost1.equivalent\endcsname{0.0}
\expandafter\def\csname val@rev.desc.d1.uiv.p1.C.center-d1.tau-source.tost1.p\endcsname{0.970}
\expandafter\def\csname cnt@rev.desc.d1.uiv.p1.C.center-d1.tau-source.tost1.p\endcsname{1}
\expandafter\def\csname pct@rev.desc.d1.uiv.p1.C.center-d1.tau-source.tost1.p\endcsname{97.0}
\expandafter\def\csname ppt@rev.desc.d1.uiv.p1.C.center-d1.tau-source.tost1.p\endcsname{$+$97.0}
\expandafter\def\csname val@rev.desc.d1.uiv.p1.C.center-d1.tau-source.tost5.ci90-inside\endcsname{0}
\expandafter\def\csname cnt@rev.desc.d1.uiv.p1.C.center-d1.tau-source.tost5.ci90-inside\endcsname{0}
\expandafter\def\csname pct@rev.desc.d1.uiv.p1.C.center-d1.tau-source.tost5.ci90-inside\endcsname{0.0}
\expandafter\def\csname ppt@rev.desc.d1.uiv.p1.C.center-d1.tau-source.tost5.ci90-inside\endcsname{0.0}
\expandafter\def\csname val@rev.desc.d1.uiv.p1.C.center-d1.tau-source.tost5.equivalent\endcsname{0}
\expandafter\def\csname cnt@rev.desc.d1.uiv.p1.C.center-d1.tau-source.tost5.equivalent\endcsname{0}
\expandafter\def\csname pct@rev.desc.d1.uiv.p1.C.center-d1.tau-source.tost5.equivalent\endcsname{0.0}
\expandafter\def\csname ppt@rev.desc.d1.uiv.p1.C.center-d1.tau-source.tost5.equivalent\endcsname{0.0}
\expandafter\def\csname val@rev.desc.d1.uiv.p1.C.center-d1.tau-source.tost5.p\endcsname{0.753}
\expandafter\def\csname cnt@rev.desc.d1.uiv.p1.C.center-d1.tau-source.tost5.p\endcsname{1}
\expandafter\def\csname pct@rev.desc.d1.uiv.p1.C.center-d1.tau-source.tost5.p\endcsname{75.3}
\expandafter\def\csname ppt@rev.desc.d1.uiv.p1.C.center-d1.tau-source.tost5.p\endcsname{$+$75.3}
\expandafter\def\csname val@rev.desc.d1.uiv.p1.C.center-d1.tau-target.ci90.0\endcsname{-0.062}
\expandafter\def\csname cnt@rev.desc.d1.uiv.p1.C.center-d1.tau-target.ci90.0\endcsname{0}
\expandafter\def\csname pct@rev.desc.d1.uiv.p1.C.center-d1.tau-target.ci90.0\endcsname{$-$6.2}
\expandafter\def\csname ppt@rev.desc.d1.uiv.p1.C.center-d1.tau-target.ci90.0\endcsname{$-$6.2}
\expandafter\def\csname val@rev.desc.d1.uiv.p1.C.center-d1.tau-target.ci90.1\endcsname{0.073}
\expandafter\def\csname cnt@rev.desc.d1.uiv.p1.C.center-d1.tau-target.ci90.1\endcsname{0}
\expandafter\def\csname pct@rev.desc.d1.uiv.p1.C.center-d1.tau-target.ci90.1\endcsname{7.3}
\expandafter\def\csname ppt@rev.desc.d1.uiv.p1.C.center-d1.tau-target.ci90.1\endcsname{$+$7.3}
\expandafter\def\csname val@rev.desc.d1.uiv.p1.C.center-d1.tau-target.ci95-excludes-0\endcsname{0}
\expandafter\def\csname cnt@rev.desc.d1.uiv.p1.C.center-d1.tau-target.ci95-excludes-0\endcsname{0}
\expandafter\def\csname pct@rev.desc.d1.uiv.p1.C.center-d1.tau-target.ci95-excludes-0\endcsname{0.0}
\expandafter\def\csname ppt@rev.desc.d1.uiv.p1.C.center-d1.tau-target.ci95-excludes-0\endcsname{0.0}
\expandafter\def\csname val@rev.desc.d1.uiv.p1.C.center-d1.tau-target.ci95.0\endcsname{-0.073}
\expandafter\def\csname cnt@rev.desc.d1.uiv.p1.C.center-d1.tau-target.ci95.0\endcsname{0}
\expandafter\def\csname pct@rev.desc.d1.uiv.p1.C.center-d1.tau-target.ci95.0\endcsname{$-$7.3}
\expandafter\def\csname ppt@rev.desc.d1.uiv.p1.C.center-d1.tau-target.ci95.0\endcsname{$-$7.3}
\expandafter\def\csname val@rev.desc.d1.uiv.p1.C.center-d1.tau-target.ci95.1\endcsname{0.083}
\expandafter\def\csname cnt@rev.desc.d1.uiv.p1.C.center-d1.tau-target.ci95.1\endcsname{0}
\expandafter\def\csname pct@rev.desc.d1.uiv.p1.C.center-d1.tau-target.ci95.1\endcsname{8.3}
\expandafter\def\csname ppt@rev.desc.d1.uiv.p1.C.center-d1.tau-target.ci95.1\endcsname{$+$8.3}
\expandafter\def\csname val@rev.desc.d1.uiv.p1.C.center-d1.tau-target.mean\endcsname{0.010}
\expandafter\def\csname cnt@rev.desc.d1.uiv.p1.C.center-d1.tau-target.mean\endcsname{0}
\expandafter\def\csname pct@rev.desc.d1.uiv.p1.C.center-d1.tau-target.mean\endcsname{1.0}
\expandafter\def\csname ppt@rev.desc.d1.uiv.p1.C.center-d1.tau-target.mean\endcsname{$+$1.0}
\expandafter\def\csname val@rev.desc.d1.uiv.p1.C.center-d1.tau-target.n-apps\endcsname{12}
\expandafter\def\csname cnt@rev.desc.d1.uiv.p1.C.center-d1.tau-target.n-apps\endcsname{12}
\expandafter\def\csname pct@rev.desc.d1.uiv.p1.C.center-d1.tau-target.n-apps\endcsname{1200.0}
\expandafter\def\csname ppt@rev.desc.d1.uiv.p1.C.center-d1.tau-target.n-apps\endcsname{$+$1200.0}
\expandafter\def\csname val@rev.desc.d1.uiv.p1.C.center-d1.tau-target.tost1.ci90-inside\endcsname{0}
\expandafter\def\csname cnt@rev.desc.d1.uiv.p1.C.center-d1.tau-target.tost1.ci90-inside\endcsname{0}
\expandafter\def\csname pct@rev.desc.d1.uiv.p1.C.center-d1.tau-target.tost1.ci90-inside\endcsname{0.0}
\expandafter\def\csname ppt@rev.desc.d1.uiv.p1.C.center-d1.tau-target.tost1.ci90-inside\endcsname{0.0}
\expandafter\def\csname val@rev.desc.d1.uiv.p1.C.center-d1.tau-target.tost1.equivalent\endcsname{0}
\expandafter\def\csname cnt@rev.desc.d1.uiv.p1.C.center-d1.tau-target.tost1.equivalent\endcsname{0}
\expandafter\def\csname pct@rev.desc.d1.uiv.p1.C.center-d1.tau-target.tost1.equivalent\endcsname{0.0}
\expandafter\def\csname ppt@rev.desc.d1.uiv.p1.C.center-d1.tau-target.tost1.equivalent\endcsname{0.0}
\expandafter\def\csname val@rev.desc.d1.uiv.p1.C.center-d1.tau-target.tost1.p\endcsname{0.549}
\expandafter\def\csname cnt@rev.desc.d1.uiv.p1.C.center-d1.tau-target.tost1.p\endcsname{1}
\expandafter\def\csname pct@rev.desc.d1.uiv.p1.C.center-d1.tau-target.tost1.p\endcsname{54.9}
\expandafter\def\csname ppt@rev.desc.d1.uiv.p1.C.center-d1.tau-target.tost1.p\endcsname{$+$54.9}
\expandafter\def\csname val@rev.desc.d1.uiv.p1.C.center-d1.tau-target.tost5.ci90-inside\endcsname{0}
\expandafter\def\csname cnt@rev.desc.d1.uiv.p1.C.center-d1.tau-target.tost5.ci90-inside\endcsname{0}
\expandafter\def\csname pct@rev.desc.d1.uiv.p1.C.center-d1.tau-target.tost5.ci90-inside\endcsname{0.0}
\expandafter\def\csname ppt@rev.desc.d1.uiv.p1.C.center-d1.tau-target.tost5.ci90-inside\endcsname{0.0}
\expandafter\def\csname val@rev.desc.d1.uiv.p1.C.center-d1.tau-target.tost5.equivalent\endcsname{0}
\expandafter\def\csname cnt@rev.desc.d1.uiv.p1.C.center-d1.tau-target.tost5.equivalent\endcsname{0}
\expandafter\def\csname pct@rev.desc.d1.uiv.p1.C.center-d1.tau-target.tost5.equivalent\endcsname{0.0}
\expandafter\def\csname ppt@rev.desc.d1.uiv.p1.C.center-d1.tau-target.tost5.equivalent\endcsname{0.0}
\expandafter\def\csname val@rev.desc.d1.uiv.p1.C.center-d1.tau-target.tost5.p\endcsname{0.185}
\expandafter\def\csname cnt@rev.desc.d1.uiv.p1.C.center-d1.tau-target.tost5.p\endcsname{0}
\expandafter\def\csname pct@rev.desc.d1.uiv.p1.C.center-d1.tau-target.tost5.p\endcsname{18.5}
\expandafter\def\csname ppt@rev.desc.d1.uiv.p1.C.center-d1.tau-target.tost5.p\endcsname{$+$18.5}
\expandafter\def\csname val@rev.desc.d1.uiv.p1.C.center-d1.zero-estimate\endcsname{0}
\expandafter\def\csname cnt@rev.desc.d1.uiv.p1.C.center-d1.zero-estimate\endcsname{0}
\expandafter\def\csname pct@rev.desc.d1.uiv.p1.C.center-d1.zero-estimate\endcsname{0.0}
\expandafter\def\csname ppt@rev.desc.d1.uiv.p1.C.center-d1.zero-estimate\endcsname{0.0}
\expandafter\def\csname val@rev.desc.d1.uiv.p1.C.context-d1.delta-tau.ci90.0\endcsname{-0.031}
\expandafter\def\csname cnt@rev.desc.d1.uiv.p1.C.context-d1.delta-tau.ci90.0\endcsname{0}
\expandafter\def\csname pct@rev.desc.d1.uiv.p1.C.context-d1.delta-tau.ci90.0\endcsname{$-$3.1}
\expandafter\def\csname ppt@rev.desc.d1.uiv.p1.C.context-d1.delta-tau.ci90.0\endcsname{$-$3.1}
\expandafter\def\csname val@rev.desc.d1.uiv.p1.C.context-d1.delta-tau.ci90.1\endcsname{0.125}
\expandafter\def\csname cnt@rev.desc.d1.uiv.p1.C.context-d1.delta-tau.ci90.1\endcsname{0}
\expandafter\def\csname pct@rev.desc.d1.uiv.p1.C.context-d1.delta-tau.ci90.1\endcsname{12.5}
\expandafter\def\csname ppt@rev.desc.d1.uiv.p1.C.context-d1.delta-tau.ci90.1\endcsname{$+$12.5}
\expandafter\def\csname val@rev.desc.d1.uiv.p1.C.context-d1.delta-tau.ci95-excludes-0\endcsname{0}
\expandafter\def\csname cnt@rev.desc.d1.uiv.p1.C.context-d1.delta-tau.ci95-excludes-0\endcsname{0}
\expandafter\def\csname pct@rev.desc.d1.uiv.p1.C.context-d1.delta-tau.ci95-excludes-0\endcsname{0.0}
\expandafter\def\csname ppt@rev.desc.d1.uiv.p1.C.context-d1.delta-tau.ci95-excludes-0\endcsname{0.0}
\expandafter\def\csname val@rev.desc.d1.uiv.p1.C.context-d1.delta-tau.ci95.0\endcsname{-0.052}
\expandafter\def\csname cnt@rev.desc.d1.uiv.p1.C.context-d1.delta-tau.ci95.0\endcsname{0}
\expandafter\def\csname pct@rev.desc.d1.uiv.p1.C.context-d1.delta-tau.ci95.0\endcsname{$-$5.2}
\expandafter\def\csname ppt@rev.desc.d1.uiv.p1.C.context-d1.delta-tau.ci95.0\endcsname{$-$5.2}
\expandafter\def\csname val@rev.desc.d1.uiv.p1.C.context-d1.delta-tau.ci95.1\endcsname{0.146}
\expandafter\def\csname cnt@rev.desc.d1.uiv.p1.C.context-d1.delta-tau.ci95.1\endcsname{0}
\expandafter\def\csname pct@rev.desc.d1.uiv.p1.C.context-d1.delta-tau.ci95.1\endcsname{14.6}
\expandafter\def\csname ppt@rev.desc.d1.uiv.p1.C.context-d1.delta-tau.ci95.1\endcsname{$+$14.6}
\expandafter\def\csname val@rev.desc.d1.uiv.p1.C.context-d1.delta-tau.mean\endcsname{0.052}
\expandafter\def\csname cnt@rev.desc.d1.uiv.p1.C.context-d1.delta-tau.mean\endcsname{0}
\expandafter\def\csname pct@rev.desc.d1.uiv.p1.C.context-d1.delta-tau.mean\endcsname{5.2}
\expandafter\def\csname ppt@rev.desc.d1.uiv.p1.C.context-d1.delta-tau.mean\endcsname{$+$5.2}
\expandafter\def\csname val@rev.desc.d1.uiv.p1.C.context-d1.delta-tau.n-apps\endcsname{12}
\expandafter\def\csname cnt@rev.desc.d1.uiv.p1.C.context-d1.delta-tau.n-apps\endcsname{12}
\expandafter\def\csname pct@rev.desc.d1.uiv.p1.C.context-d1.delta-tau.n-apps\endcsname{1200.0}
\expandafter\def\csname ppt@rev.desc.d1.uiv.p1.C.context-d1.delta-tau.n-apps\endcsname{$+$1200.0}
\expandafter\def\csname val@rev.desc.d1.uiv.p1.C.context-d1.delta-tau.tost1.ci90-inside\endcsname{0}
\expandafter\def\csname cnt@rev.desc.d1.uiv.p1.C.context-d1.delta-tau.tost1.ci90-inside\endcsname{0}
\expandafter\def\csname pct@rev.desc.d1.uiv.p1.C.context-d1.delta-tau.tost1.ci90-inside\endcsname{0.0}
\expandafter\def\csname ppt@rev.desc.d1.uiv.p1.C.context-d1.delta-tau.tost1.ci90-inside\endcsname{0.0}
\expandafter\def\csname val@rev.desc.d1.uiv.p1.C.context-d1.delta-tau.tost1.equivalent\endcsname{0}
\expandafter\def\csname cnt@rev.desc.d1.uiv.p1.C.context-d1.delta-tau.tost1.equivalent\endcsname{0}
\expandafter\def\csname pct@rev.desc.d1.uiv.p1.C.context-d1.delta-tau.tost1.equivalent\endcsname{0.0}
\expandafter\def\csname ppt@rev.desc.d1.uiv.p1.C.context-d1.delta-tau.tost1.equivalent\endcsname{0.0}
\expandafter\def\csname val@rev.desc.d1.uiv.p1.C.context-d1.delta-tau.tost1.p\endcsname{0.817}
\expandafter\def\csname cnt@rev.desc.d1.uiv.p1.C.context-d1.delta-tau.tost1.p\endcsname{1}
\expandafter\def\csname pct@rev.desc.d1.uiv.p1.C.context-d1.delta-tau.tost1.p\endcsname{81.7}
\expandafter\def\csname ppt@rev.desc.d1.uiv.p1.C.context-d1.delta-tau.tost1.p\endcsname{$+$81.7}
\expandafter\def\csname val@rev.desc.d1.uiv.p1.C.context-d1.delta-tau.tost5.ci90-inside\endcsname{0}
\expandafter\def\csname cnt@rev.desc.d1.uiv.p1.C.context-d1.delta-tau.tost5.ci90-inside\endcsname{0}
\expandafter\def\csname pct@rev.desc.d1.uiv.p1.C.context-d1.delta-tau.tost5.ci90-inside\endcsname{0.0}
\expandafter\def\csname ppt@rev.desc.d1.uiv.p1.C.context-d1.delta-tau.tost5.ci90-inside\endcsname{0.0}
\expandafter\def\csname val@rev.desc.d1.uiv.p1.C.context-d1.delta-tau.tost5.equivalent\endcsname{0}
\expandafter\def\csname cnt@rev.desc.d1.uiv.p1.C.context-d1.delta-tau.tost5.equivalent\endcsname{0}
\expandafter\def\csname pct@rev.desc.d1.uiv.p1.C.context-d1.delta-tau.tost5.equivalent\endcsname{0.0}
\expandafter\def\csname ppt@rev.desc.d1.uiv.p1.C.context-d1.delta-tau.tost5.equivalent\endcsname{0.0}
\expandafter\def\csname val@rev.desc.d1.uiv.p1.C.context-d1.delta-tau.tost5.p\endcsname{0.537}
\expandafter\def\csname cnt@rev.desc.d1.uiv.p1.C.context-d1.delta-tau.tost5.p\endcsname{1}
\expandafter\def\csname pct@rev.desc.d1.uiv.p1.C.context-d1.delta-tau.tost5.p\endcsname{53.7}
\expandafter\def\csname ppt@rev.desc.d1.uiv.p1.C.context-d1.delta-tau.tost5.p\endcsname{$+$53.7}
\expandafter\def\csname val@rev.desc.d1.uiv.p1.C.context-d1.material-flip\endcsname{0}
\expandafter\def\csname cnt@rev.desc.d1.uiv.p1.C.context-d1.material-flip\endcsname{0}
\expandafter\def\csname pct@rev.desc.d1.uiv.p1.C.context-d1.material-flip\endcsname{0.0}
\expandafter\def\csname ppt@rev.desc.d1.uiv.p1.C.context-d1.material-flip\endcsname{0.0}
\expandafter\def\csname val@rev.desc.d1.uiv.p1.C.context-d1.n-apps\endcsname{12}
\expandafter\def\csname cnt@rev.desc.d1.uiv.p1.C.context-d1.n-apps\endcsname{12}
\expandafter\def\csname pct@rev.desc.d1.uiv.p1.C.context-d1.n-apps\endcsname{1200.0}
\expandafter\def\csname ppt@rev.desc.d1.uiv.p1.C.context-d1.n-apps\endcsname{$+$1200.0}
\expandafter\def\csname val@rev.desc.d1.uiv.p1.C.context-d1.n-tasks\endcsname{96}
\expandafter\def\csname cnt@rev.desc.d1.uiv.p1.C.context-d1.n-tasks\endcsname{96}
\expandafter\def\csname pct@rev.desc.d1.uiv.p1.C.context-d1.n-tasks\endcsname{9600.0}
\expandafter\def\csname ppt@rev.desc.d1.uiv.p1.C.context-d1.n-tasks\endcsname{$+$9600.0}
\expandafter\def\csname val@rev.desc.d1.uiv.p1.C.context-d1.sign-differs\endcsname{0}
\expandafter\def\csname cnt@rev.desc.d1.uiv.p1.C.context-d1.sign-differs\endcsname{0}
\expandafter\def\csname pct@rev.desc.d1.uiv.p1.C.context-d1.sign-differs\endcsname{0.0}
\expandafter\def\csname ppt@rev.desc.d1.uiv.p1.C.context-d1.sign-differs\endcsname{0.0}
\expandafter\def\csname val@rev.desc.d1.uiv.p1.C.context-d1.tau-source.ci90.0\endcsname{-0.135}
\expandafter\def\csname cnt@rev.desc.d1.uiv.p1.C.context-d1.tau-source.ci90.0\endcsname{0}
\expandafter\def\csname pct@rev.desc.d1.uiv.p1.C.context-d1.tau-source.ci90.0\endcsname{$-$13.5}
\expandafter\def\csname ppt@rev.desc.d1.uiv.p1.C.context-d1.tau-source.ci90.0\endcsname{$-$13.5}
\expandafter\def\csname val@rev.desc.d1.uiv.p1.C.context-d1.tau-source.ci90.1\endcsname{-0.021}
\expandafter\def\csname cnt@rev.desc.d1.uiv.p1.C.context-d1.tau-source.ci90.1\endcsname{0}
\expandafter\def\csname pct@rev.desc.d1.uiv.p1.C.context-d1.tau-source.ci90.1\endcsname{$-$2.1}
\expandafter\def\csname ppt@rev.desc.d1.uiv.p1.C.context-d1.tau-source.ci90.1\endcsname{$-$2.1}
\expandafter\def\csname val@rev.desc.d1.uiv.p1.C.context-d1.tau-source.ci95-excludes-0\endcsname{1}
\expandafter\def\csname cnt@rev.desc.d1.uiv.p1.C.context-d1.tau-source.ci95-excludes-0\endcsname{1}
\expandafter\def\csname pct@rev.desc.d1.uiv.p1.C.context-d1.tau-source.ci95-excludes-0\endcsname{100.0}
\expandafter\def\csname ppt@rev.desc.d1.uiv.p1.C.context-d1.tau-source.ci95-excludes-0\endcsname{$+$100.0}
\expandafter\def\csname val@rev.desc.d1.uiv.p1.C.context-d1.tau-source.ci95.0\endcsname{-0.146}
\expandafter\def\csname cnt@rev.desc.d1.uiv.p1.C.context-d1.tau-source.ci95.0\endcsname{0}
\expandafter\def\csname pct@rev.desc.d1.uiv.p1.C.context-d1.tau-source.ci95.0\endcsname{$-$14.6}
\expandafter\def\csname ppt@rev.desc.d1.uiv.p1.C.context-d1.tau-source.ci95.0\endcsname{$-$14.6}
\expandafter\def\csname val@rev.desc.d1.uiv.p1.C.context-d1.tau-source.ci95.1\endcsname{-0.010}
\expandafter\def\csname cnt@rev.desc.d1.uiv.p1.C.context-d1.tau-source.ci95.1\endcsname{0}
\expandafter\def\csname pct@rev.desc.d1.uiv.p1.C.context-d1.tau-source.ci95.1\endcsname{$-$1.0}
\expandafter\def\csname ppt@rev.desc.d1.uiv.p1.C.context-d1.tau-source.ci95.1\endcsname{$-$1.0}
\expandafter\def\csname val@rev.desc.d1.uiv.p1.C.context-d1.tau-source.mean\endcsname{-0.073}
\expandafter\def\csname cnt@rev.desc.d1.uiv.p1.C.context-d1.tau-source.mean\endcsname{0}
\expandafter\def\csname pct@rev.desc.d1.uiv.p1.C.context-d1.tau-source.mean\endcsname{$-$7.3}
\expandafter\def\csname ppt@rev.desc.d1.uiv.p1.C.context-d1.tau-source.mean\endcsname{$-$7.3}
\expandafter\def\csname val@rev.desc.d1.uiv.p1.C.context-d1.tau-source.n-apps\endcsname{12}
\expandafter\def\csname cnt@rev.desc.d1.uiv.p1.C.context-d1.tau-source.n-apps\endcsname{12}
\expandafter\def\csname pct@rev.desc.d1.uiv.p1.C.context-d1.tau-source.n-apps\endcsname{1200.0}
\expandafter\def\csname ppt@rev.desc.d1.uiv.p1.C.context-d1.tau-source.n-apps\endcsname{$+$1200.0}
\expandafter\def\csname val@rev.desc.d1.uiv.p1.C.context-d1.tau-source.tost1.ci90-inside\endcsname{0}
\expandafter\def\csname cnt@rev.desc.d1.uiv.p1.C.context-d1.tau-source.tost1.ci90-inside\endcsname{0}
\expandafter\def\csname pct@rev.desc.d1.uiv.p1.C.context-d1.tau-source.tost1.ci90-inside\endcsname{0.0}
\expandafter\def\csname ppt@rev.desc.d1.uiv.p1.C.context-d1.tau-source.tost1.ci90-inside\endcsname{0.0}
\expandafter\def\csname val@rev.desc.d1.uiv.p1.C.context-d1.tau-source.tost1.equivalent\endcsname{0}
\expandafter\def\csname cnt@rev.desc.d1.uiv.p1.C.context-d1.tau-source.tost1.equivalent\endcsname{0}
\expandafter\def\csname pct@rev.desc.d1.uiv.p1.C.context-d1.tau-source.tost1.equivalent\endcsname{0.0}
\expandafter\def\csname ppt@rev.desc.d1.uiv.p1.C.context-d1.tau-source.tost1.equivalent\endcsname{0.0}
\expandafter\def\csname val@rev.desc.d1.uiv.p1.C.context-d1.tau-source.tost1.p\endcsname{0.976}
\expandafter\def\csname cnt@rev.desc.d1.uiv.p1.C.context-d1.tau-source.tost1.p\endcsname{1}
\expandafter\def\csname pct@rev.desc.d1.uiv.p1.C.context-d1.tau-source.tost1.p\endcsname{97.6}
\expandafter\def\csname ppt@rev.desc.d1.uiv.p1.C.context-d1.tau-source.tost1.p\endcsname{$+$97.6}
\expandafter\def\csname val@rev.desc.d1.uiv.p1.C.context-d1.tau-source.tost5.ci90-inside\endcsname{0}
\expandafter\def\csname cnt@rev.desc.d1.uiv.p1.C.context-d1.tau-source.tost5.ci90-inside\endcsname{0}
\expandafter\def\csname pct@rev.desc.d1.uiv.p1.C.context-d1.tau-source.tost5.ci90-inside\endcsname{0.0}
\expandafter\def\csname ppt@rev.desc.d1.uiv.p1.C.context-d1.tau-source.tost5.ci90-inside\endcsname{0.0}
\expandafter\def\csname val@rev.desc.d1.uiv.p1.C.context-d1.tau-source.tost5.equivalent\endcsname{0}
\expandafter\def\csname cnt@rev.desc.d1.uiv.p1.C.context-d1.tau-source.tost5.equivalent\endcsname{0}
\expandafter\def\csname pct@rev.desc.d1.uiv.p1.C.context-d1.tau-source.tost5.equivalent\endcsname{0.0}
\expandafter\def\csname ppt@rev.desc.d1.uiv.p1.C.context-d1.tau-source.tost5.equivalent\endcsname{0.0}
\expandafter\def\csname val@rev.desc.d1.uiv.p1.C.context-d1.tau-source.tost5.p\endcsname{0.763}
\expandafter\def\csname cnt@rev.desc.d1.uiv.p1.C.context-d1.tau-source.tost5.p\endcsname{1}
\expandafter\def\csname pct@rev.desc.d1.uiv.p1.C.context-d1.tau-source.tost5.p\endcsname{76.3}
\expandafter\def\csname ppt@rev.desc.d1.uiv.p1.C.context-d1.tau-source.tost5.p\endcsname{$+$76.3}
\expandafter\def\csname val@rev.desc.d1.uiv.p1.C.context-d1.tau-target.ci90.0\endcsname{-0.083}
\expandafter\def\csname cnt@rev.desc.d1.uiv.p1.C.context-d1.tau-target.ci90.0\endcsname{0}
\expandafter\def\csname pct@rev.desc.d1.uiv.p1.C.context-d1.tau-target.ci90.0\endcsname{$-$8.3}
\expandafter\def\csname ppt@rev.desc.d1.uiv.p1.C.context-d1.tau-target.ci90.0\endcsname{$-$8.3}
\expandafter\def\csname val@rev.desc.d1.uiv.p1.C.context-d1.tau-target.ci90.1\endcsname{0.042}
\expandafter\def\csname cnt@rev.desc.d1.uiv.p1.C.context-d1.tau-target.ci90.1\endcsname{0}
\expandafter\def\csname pct@rev.desc.d1.uiv.p1.C.context-d1.tau-target.ci90.1\endcsname{4.2}
\expandafter\def\csname ppt@rev.desc.d1.uiv.p1.C.context-d1.tau-target.ci90.1\endcsname{$+$4.2}
\expandafter\def\csname val@rev.desc.d1.uiv.p1.C.context-d1.tau-target.ci95-excludes-0\endcsname{0}
\expandafter\def\csname cnt@rev.desc.d1.uiv.p1.C.context-d1.tau-target.ci95-excludes-0\endcsname{0}
\expandafter\def\csname pct@rev.desc.d1.uiv.p1.C.context-d1.tau-target.ci95-excludes-0\endcsname{0.0}
\expandafter\def\csname ppt@rev.desc.d1.uiv.p1.C.context-d1.tau-target.ci95-excludes-0\endcsname{0.0}
\expandafter\def\csname val@rev.desc.d1.uiv.p1.C.context-d1.tau-target.ci95.0\endcsname{-0.104}
\expandafter\def\csname cnt@rev.desc.d1.uiv.p1.C.context-d1.tau-target.ci95.0\endcsname{0}
\expandafter\def\csname pct@rev.desc.d1.uiv.p1.C.context-d1.tau-target.ci95.0\endcsname{$-$10.4}
\expandafter\def\csname ppt@rev.desc.d1.uiv.p1.C.context-d1.tau-target.ci95.0\endcsname{$-$10.4}
\expandafter\def\csname val@rev.desc.d1.uiv.p1.C.context-d1.tau-target.ci95.1\endcsname{0.052}
\expandafter\def\csname cnt@rev.desc.d1.uiv.p1.C.context-d1.tau-target.ci95.1\endcsname{0}
\expandafter\def\csname pct@rev.desc.d1.uiv.p1.C.context-d1.tau-target.ci95.1\endcsname{5.2}
\expandafter\def\csname ppt@rev.desc.d1.uiv.p1.C.context-d1.tau-target.ci95.1\endcsname{$+$5.2}
\expandafter\def\csname val@rev.desc.d1.uiv.p1.C.context-d1.tau-target.mean\endcsname{-0.021}
\expandafter\def\csname cnt@rev.desc.d1.uiv.p1.C.context-d1.tau-target.mean\endcsname{0}
\expandafter\def\csname pct@rev.desc.d1.uiv.p1.C.context-d1.tau-target.mean\endcsname{$-$2.1}
\expandafter\def\csname ppt@rev.desc.d1.uiv.p1.C.context-d1.tau-target.mean\endcsname{$-$2.1}
\expandafter\def\csname val@rev.desc.d1.uiv.p1.C.context-d1.tau-target.n-apps\endcsname{12}
\expandafter\def\csname cnt@rev.desc.d1.uiv.p1.C.context-d1.tau-target.n-apps\endcsname{12}
\expandafter\def\csname pct@rev.desc.d1.uiv.p1.C.context-d1.tau-target.n-apps\endcsname{1200.0}
\expandafter\def\csname ppt@rev.desc.d1.uiv.p1.C.context-d1.tau-target.n-apps\endcsname{$+$1200.0}
\expandafter\def\csname val@rev.desc.d1.uiv.p1.C.context-d1.tau-target.tost1.ci90-inside\endcsname{0}
\expandafter\def\csname cnt@rev.desc.d1.uiv.p1.C.context-d1.tau-target.tost1.ci90-inside\endcsname{0}
\expandafter\def\csname pct@rev.desc.d1.uiv.p1.C.context-d1.tau-target.tost1.ci90-inside\endcsname{0.0}
\expandafter\def\csname ppt@rev.desc.d1.uiv.p1.C.context-d1.tau-target.tost1.ci90-inside\endcsname{0.0}
\expandafter\def\csname val@rev.desc.d1.uiv.p1.C.context-d1.tau-target.tost1.equivalent\endcsname{0}
\expandafter\def\csname cnt@rev.desc.d1.uiv.p1.C.context-d1.tau-target.tost1.equivalent\endcsname{0}
\expandafter\def\csname pct@rev.desc.d1.uiv.p1.C.context-d1.tau-target.tost1.equivalent\endcsname{0.0}
\expandafter\def\csname ppt@rev.desc.d1.uiv.p1.C.context-d1.tau-target.tost1.equivalent\endcsname{0.0}
\expandafter\def\csname val@rev.desc.d1.uiv.p1.C.context-d1.tau-target.tost1.p\endcsname{0.639}
\expandafter\def\csname cnt@rev.desc.d1.uiv.p1.C.context-d1.tau-target.tost1.p\endcsname{1}
\expandafter\def\csname pct@rev.desc.d1.uiv.p1.C.context-d1.tau-target.tost1.p\endcsname{63.9}
\expandafter\def\csname ppt@rev.desc.d1.uiv.p1.C.context-d1.tau-target.tost1.p\endcsname{$+$63.9}
\expandafter\def\csname val@rev.desc.d1.uiv.p1.C.context-d1.tau-target.tost5.ci90-inside\endcsname{0}
\expandafter\def\csname cnt@rev.desc.d1.uiv.p1.C.context-d1.tau-target.tost5.ci90-inside\endcsname{0}
\expandafter\def\csname pct@rev.desc.d1.uiv.p1.C.context-d1.tau-target.tost5.ci90-inside\endcsname{0.0}
\expandafter\def\csname ppt@rev.desc.d1.uiv.p1.C.context-d1.tau-target.tost5.ci90-inside\endcsname{0.0}
\expandafter\def\csname val@rev.desc.d1.uiv.p1.C.context-d1.tau-target.tost5.equivalent\endcsname{0}
\expandafter\def\csname cnt@rev.desc.d1.uiv.p1.C.context-d1.tau-target.tost5.equivalent\endcsname{0}
\expandafter\def\csname pct@rev.desc.d1.uiv.p1.C.context-d1.tau-target.tost5.equivalent\endcsname{0.0}
\expandafter\def\csname ppt@rev.desc.d1.uiv.p1.C.context-d1.tau-target.tost5.equivalent\endcsname{0.0}
\expandafter\def\csname val@rev.desc.d1.uiv.p1.C.context-d1.tau-target.tost5.p\endcsname{0.247}
\expandafter\def\csname cnt@rev.desc.d1.uiv.p1.C.context-d1.tau-target.tost5.p\endcsname{0}
\expandafter\def\csname pct@rev.desc.d1.uiv.p1.C.context-d1.tau-target.tost5.p\endcsname{24.7}
\expandafter\def\csname ppt@rev.desc.d1.uiv.p1.C.context-d1.tau-target.tost5.p\endcsname{$+$24.7}
\expandafter\def\csname val@rev.desc.d1.uiv.p1.C.context-d1.zero-estimate\endcsname{0}
\expandafter\def\csname cnt@rev.desc.d1.uiv.p1.C.context-d1.zero-estimate\endcsname{0}
\expandafter\def\csname pct@rev.desc.d1.uiv.p1.C.context-d1.zero-estimate\endcsname{0.0}
\expandafter\def\csname ppt@rev.desc.d1.uiv.p1.C.context-d1.zero-estimate\endcsname{0.0}
\expandafter\def\csname val@rev.desc.d1.uiv.p1.C.dual-d2.delta-tau.ci90.0\endcsname{-0.021}
\expandafter\def\csname cnt@rev.desc.d1.uiv.p1.C.dual-d2.delta-tau.ci90.0\endcsname{0}
\expandafter\def\csname pct@rev.desc.d1.uiv.p1.C.dual-d2.delta-tau.ci90.0\endcsname{$-$2.1}
\expandafter\def\csname ppt@rev.desc.d1.uiv.p1.C.dual-d2.delta-tau.ci90.0\endcsname{$-$2.1}
\expandafter\def\csname val@rev.desc.d1.uiv.p1.C.dual-d2.delta-tau.ci90.1\endcsname{0.177}
\expandafter\def\csname cnt@rev.desc.d1.uiv.p1.C.dual-d2.delta-tau.ci90.1\endcsname{0}
\expandafter\def\csname pct@rev.desc.d1.uiv.p1.C.dual-d2.delta-tau.ci90.1\endcsname{17.7}
\expandafter\def\csname ppt@rev.desc.d1.uiv.p1.C.dual-d2.delta-tau.ci90.1\endcsname{$+$17.7}
\expandafter\def\csname val@rev.desc.d1.uiv.p1.C.dual-d2.delta-tau.ci95-excludes-0\endcsname{0}
\expandafter\def\csname cnt@rev.desc.d1.uiv.p1.C.dual-d2.delta-tau.ci95-excludes-0\endcsname{0}
\expandafter\def\csname pct@rev.desc.d1.uiv.p1.C.dual-d2.delta-tau.ci95-excludes-0\endcsname{0.0}
\expandafter\def\csname ppt@rev.desc.d1.uiv.p1.C.dual-d2.delta-tau.ci95-excludes-0\endcsname{0.0}
\expandafter\def\csname val@rev.desc.d1.uiv.p1.C.dual-d2.delta-tau.ci95.0\endcsname{-0.042}
\expandafter\def\csname cnt@rev.desc.d1.uiv.p1.C.dual-d2.delta-tau.ci95.0\endcsname{0}
\expandafter\def\csname pct@rev.desc.d1.uiv.p1.C.dual-d2.delta-tau.ci95.0\endcsname{$-$4.2}
\expandafter\def\csname ppt@rev.desc.d1.uiv.p1.C.dual-d2.delta-tau.ci95.0\endcsname{$-$4.2}
\expandafter\def\csname val@rev.desc.d1.uiv.p1.C.dual-d2.delta-tau.ci95.1\endcsname{0.198}
\expandafter\def\csname cnt@rev.desc.d1.uiv.p1.C.dual-d2.delta-tau.ci95.1\endcsname{0}
\expandafter\def\csname pct@rev.desc.d1.uiv.p1.C.dual-d2.delta-tau.ci95.1\endcsname{19.8}
\expandafter\def\csname ppt@rev.desc.d1.uiv.p1.C.dual-d2.delta-tau.ci95.1\endcsname{$+$19.8}
\expandafter\def\csname val@rev.desc.d1.uiv.p1.C.dual-d2.delta-tau.mean\endcsname{0.083}
\expandafter\def\csname cnt@rev.desc.d1.uiv.p1.C.dual-d2.delta-tau.mean\endcsname{0}
\expandafter\def\csname pct@rev.desc.d1.uiv.p1.C.dual-d2.delta-tau.mean\endcsname{8.3}
\expandafter\def\csname ppt@rev.desc.d1.uiv.p1.C.dual-d2.delta-tau.mean\endcsname{$+$8.3}
\expandafter\def\csname val@rev.desc.d1.uiv.p1.C.dual-d2.delta-tau.n-apps\endcsname{12}
\expandafter\def\csname cnt@rev.desc.d1.uiv.p1.C.dual-d2.delta-tau.n-apps\endcsname{12}
\expandafter\def\csname pct@rev.desc.d1.uiv.p1.C.dual-d2.delta-tau.n-apps\endcsname{1200.0}
\expandafter\def\csname ppt@rev.desc.d1.uiv.p1.C.dual-d2.delta-tau.n-apps\endcsname{$+$1200.0}
\expandafter\def\csname val@rev.desc.d1.uiv.p1.C.dual-d2.delta-tau.tost1.ci90-inside\endcsname{0}
\expandafter\def\csname cnt@rev.desc.d1.uiv.p1.C.dual-d2.delta-tau.tost1.ci90-inside\endcsname{0}
\expandafter\def\csname pct@rev.desc.d1.uiv.p1.C.dual-d2.delta-tau.tost1.ci90-inside\endcsname{0.0}
\expandafter\def\csname ppt@rev.desc.d1.uiv.p1.C.dual-d2.delta-tau.tost1.ci90-inside\endcsname{0.0}
\expandafter\def\csname val@rev.desc.d1.uiv.p1.C.dual-d2.delta-tau.tost1.equivalent\endcsname{0}
\expandafter\def\csname cnt@rev.desc.d1.uiv.p1.C.dual-d2.delta-tau.tost1.equivalent\endcsname{0}
\expandafter\def\csname pct@rev.desc.d1.uiv.p1.C.dual-d2.delta-tau.tost1.equivalent\endcsname{0.0}
\expandafter\def\csname ppt@rev.desc.d1.uiv.p1.C.dual-d2.delta-tau.tost1.equivalent\endcsname{0.0}
\expandafter\def\csname val@rev.desc.d1.uiv.p1.C.dual-d2.delta-tau.tost1.p\endcsname{0.891}
\expandafter\def\csname cnt@rev.desc.d1.uiv.p1.C.dual-d2.delta-tau.tost1.p\endcsname{1}
\expandafter\def\csname pct@rev.desc.d1.uiv.p1.C.dual-d2.delta-tau.tost1.p\endcsname{89.1}
\expandafter\def\csname ppt@rev.desc.d1.uiv.p1.C.dual-d2.delta-tau.tost1.p\endcsname{$+$89.1}
\expandafter\def\csname val@rev.desc.d1.uiv.p1.C.dual-d2.delta-tau.tost5.ci90-inside\endcsname{0}
\expandafter\def\csname cnt@rev.desc.d1.uiv.p1.C.dual-d2.delta-tau.tost5.ci90-inside\endcsname{0}
\expandafter\def\csname pct@rev.desc.d1.uiv.p1.C.dual-d2.delta-tau.tost5.ci90-inside\endcsname{0.0}
\expandafter\def\csname ppt@rev.desc.d1.uiv.p1.C.dual-d2.delta-tau.tost5.ci90-inside\endcsname{0.0}
\expandafter\def\csname val@rev.desc.d1.uiv.p1.C.dual-d2.delta-tau.tost5.equivalent\endcsname{0}
\expandafter\def\csname cnt@rev.desc.d1.uiv.p1.C.dual-d2.delta-tau.tost5.equivalent\endcsname{0}
\expandafter\def\csname pct@rev.desc.d1.uiv.p1.C.dual-d2.delta-tau.tost5.equivalent\endcsname{0.0}
\expandafter\def\csname ppt@rev.desc.d1.uiv.p1.C.dual-d2.delta-tau.tost5.equivalent\endcsname{0.0}
\expandafter\def\csname val@rev.desc.d1.uiv.p1.C.dual-d2.delta-tau.tost5.p\endcsname{0.729}
\expandafter\def\csname cnt@rev.desc.d1.uiv.p1.C.dual-d2.delta-tau.tost5.p\endcsname{1}
\expandafter\def\csname pct@rev.desc.d1.uiv.p1.C.dual-d2.delta-tau.tost5.p\endcsname{72.9}
\expandafter\def\csname ppt@rev.desc.d1.uiv.p1.C.dual-d2.delta-tau.tost5.p\endcsname{$+$72.9}
\expandafter\def\csname val@rev.desc.d1.uiv.p1.C.dual-d2.material-flip\endcsname{0}
\expandafter\def\csname cnt@rev.desc.d1.uiv.p1.C.dual-d2.material-flip\endcsname{0}
\expandafter\def\csname pct@rev.desc.d1.uiv.p1.C.dual-d2.material-flip\endcsname{0.0}
\expandafter\def\csname ppt@rev.desc.d1.uiv.p1.C.dual-d2.material-flip\endcsname{0.0}
\expandafter\def\csname val@rev.desc.d1.uiv.p1.C.dual-d2.n-apps\endcsname{12}
\expandafter\def\csname cnt@rev.desc.d1.uiv.p1.C.dual-d2.n-apps\endcsname{12}
\expandafter\def\csname pct@rev.desc.d1.uiv.p1.C.dual-d2.n-apps\endcsname{1200.0}
\expandafter\def\csname ppt@rev.desc.d1.uiv.p1.C.dual-d2.n-apps\endcsname{$+$1200.0}
\expandafter\def\csname val@rev.desc.d1.uiv.p1.C.dual-d2.n-tasks\endcsname{96}
\expandafter\def\csname cnt@rev.desc.d1.uiv.p1.C.dual-d2.n-tasks\endcsname{96}
\expandafter\def\csname pct@rev.desc.d1.uiv.p1.C.dual-d2.n-tasks\endcsname{9600.0}
\expandafter\def\csname ppt@rev.desc.d1.uiv.p1.C.dual-d2.n-tasks\endcsname{$+$9600.0}
\expandafter\def\csname val@rev.desc.d1.uiv.p1.C.dual-d2.sign-differs\endcsname{0}
\expandafter\def\csname cnt@rev.desc.d1.uiv.p1.C.dual-d2.sign-differs\endcsname{0}
\expandafter\def\csname pct@rev.desc.d1.uiv.p1.C.dual-d2.sign-differs\endcsname{0.0}
\expandafter\def\csname ppt@rev.desc.d1.uiv.p1.C.dual-d2.sign-differs\endcsname{0.0}
\expandafter\def\csname val@rev.desc.d1.uiv.p1.C.dual-d2.tau-source.ci90.0\endcsname{-0.156}
\expandafter\def\csname cnt@rev.desc.d1.uiv.p1.C.dual-d2.tau-source.ci90.0\endcsname{0}
\expandafter\def\csname pct@rev.desc.d1.uiv.p1.C.dual-d2.tau-source.ci90.0\endcsname{$-$15.6}
\expandafter\def\csname ppt@rev.desc.d1.uiv.p1.C.dual-d2.tau-source.ci90.0\endcsname{$-$15.6}
\expandafter\def\csname val@rev.desc.d1.uiv.p1.C.dual-d2.tau-source.ci90.1\endcsname{0.000}
\expandafter\def\csname cnt@rev.desc.d1.uiv.p1.C.dual-d2.tau-source.ci90.1\endcsname{0}
\expandafter\def\csname pct@rev.desc.d1.uiv.p1.C.dual-d2.tau-source.ci90.1\endcsname{0.0}
\expandafter\def\csname ppt@rev.desc.d1.uiv.p1.C.dual-d2.tau-source.ci90.1\endcsname{0.0}
\expandafter\def\csname val@rev.desc.d1.uiv.p1.C.dual-d2.tau-source.ci95-excludes-0\endcsname{0}
\expandafter\def\csname cnt@rev.desc.d1.uiv.p1.C.dual-d2.tau-source.ci95-excludes-0\endcsname{0}
\expandafter\def\csname pct@rev.desc.d1.uiv.p1.C.dual-d2.tau-source.ci95-excludes-0\endcsname{0.0}
\expandafter\def\csname ppt@rev.desc.d1.uiv.p1.C.dual-d2.tau-source.ci95-excludes-0\endcsname{0.0}
\expandafter\def\csname val@rev.desc.d1.uiv.p1.C.dual-d2.tau-source.ci95.0\endcsname{-0.167}
\expandafter\def\csname cnt@rev.desc.d1.uiv.p1.C.dual-d2.tau-source.ci95.0\endcsname{0}
\expandafter\def\csname pct@rev.desc.d1.uiv.p1.C.dual-d2.tau-source.ci95.0\endcsname{$-$16.7}
\expandafter\def\csname ppt@rev.desc.d1.uiv.p1.C.dual-d2.tau-source.ci95.0\endcsname{$-$16.7}
\expandafter\def\csname val@rev.desc.d1.uiv.p1.C.dual-d2.tau-source.ci95.1\endcsname{0.021}
\expandafter\def\csname cnt@rev.desc.d1.uiv.p1.C.dual-d2.tau-source.ci95.1\endcsname{0}
\expandafter\def\csname pct@rev.desc.d1.uiv.p1.C.dual-d2.tau-source.ci95.1\endcsname{2.1}
\expandafter\def\csname ppt@rev.desc.d1.uiv.p1.C.dual-d2.tau-source.ci95.1\endcsname{$+$2.1}
\expandafter\def\csname val@rev.desc.d1.uiv.p1.C.dual-d2.tau-source.mean\endcsname{-0.083}
\expandafter\def\csname cnt@rev.desc.d1.uiv.p1.C.dual-d2.tau-source.mean\endcsname{0}
\expandafter\def\csname pct@rev.desc.d1.uiv.p1.C.dual-d2.tau-source.mean\endcsname{$-$8.3}
\expandafter\def\csname ppt@rev.desc.d1.uiv.p1.C.dual-d2.tau-source.mean\endcsname{$-$8.3}
\expandafter\def\csname val@rev.desc.d1.uiv.p1.C.dual-d2.tau-source.n-apps\endcsname{12}
\expandafter\def\csname cnt@rev.desc.d1.uiv.p1.C.dual-d2.tau-source.n-apps\endcsname{12}
\expandafter\def\csname pct@rev.desc.d1.uiv.p1.C.dual-d2.tau-source.n-apps\endcsname{1200.0}
\expandafter\def\csname ppt@rev.desc.d1.uiv.p1.C.dual-d2.tau-source.n-apps\endcsname{$+$1200.0}
\expandafter\def\csname val@rev.desc.d1.uiv.p1.C.dual-d2.tau-source.tost1.ci90-inside\endcsname{0}
\expandafter\def\csname cnt@rev.desc.d1.uiv.p1.C.dual-d2.tau-source.tost1.ci90-inside\endcsname{0}
\expandafter\def\csname pct@rev.desc.d1.uiv.p1.C.dual-d2.tau-source.tost1.ci90-inside\endcsname{0.0}
\expandafter\def\csname ppt@rev.desc.d1.uiv.p1.C.dual-d2.tau-source.tost1.ci90-inside\endcsname{0.0}
\expandafter\def\csname val@rev.desc.d1.uiv.p1.C.dual-d2.tau-source.tost1.equivalent\endcsname{0}
\expandafter\def\csname cnt@rev.desc.d1.uiv.p1.C.dual-d2.tau-source.tost1.equivalent\endcsname{0}
\expandafter\def\csname pct@rev.desc.d1.uiv.p1.C.dual-d2.tau-source.tost1.equivalent\endcsname{0.0}
\expandafter\def\csname ppt@rev.desc.d1.uiv.p1.C.dual-d2.tau-source.tost1.equivalent\endcsname{0.0}
\expandafter\def\csname val@rev.desc.d1.uiv.p1.C.dual-d2.tau-source.tost1.p\endcsname{0.946}
\expandafter\def\csname cnt@rev.desc.d1.uiv.p1.C.dual-d2.tau-source.tost1.p\endcsname{1}
\expandafter\def\csname pct@rev.desc.d1.uiv.p1.C.dual-d2.tau-source.tost1.p\endcsname{94.6}
\expandafter\def\csname ppt@rev.desc.d1.uiv.p1.C.dual-d2.tau-source.tost1.p\endcsname{$+$94.6}
\expandafter\def\csname val@rev.desc.d1.uiv.p1.C.dual-d2.tau-source.tost5.ci90-inside\endcsname{0}
\expandafter\def\csname cnt@rev.desc.d1.uiv.p1.C.dual-d2.tau-source.tost5.ci90-inside\endcsname{0}
\expandafter\def\csname pct@rev.desc.d1.uiv.p1.C.dual-d2.tau-source.tost5.ci90-inside\endcsname{0.0}
\expandafter\def\csname ppt@rev.desc.d1.uiv.p1.C.dual-d2.tau-source.tost5.ci90-inside\endcsname{0.0}
\expandafter\def\csname val@rev.desc.d1.uiv.p1.C.dual-d2.tau-source.tost5.equivalent\endcsname{0}
\expandafter\def\csname cnt@rev.desc.d1.uiv.p1.C.dual-d2.tau-source.tost5.equivalent\endcsname{0}
\expandafter\def\csname pct@rev.desc.d1.uiv.p1.C.dual-d2.tau-source.tost5.equivalent\endcsname{0.0}
\expandafter\def\csname ppt@rev.desc.d1.uiv.p1.C.dual-d2.tau-source.tost5.equivalent\endcsname{0.0}
\expandafter\def\csname val@rev.desc.d1.uiv.p1.C.dual-d2.tau-source.tost5.p\endcsname{0.790}
\expandafter\def\csname cnt@rev.desc.d1.uiv.p1.C.dual-d2.tau-source.tost5.p\endcsname{1}
\expandafter\def\csname pct@rev.desc.d1.uiv.p1.C.dual-d2.tau-source.tost5.p\endcsname{79.0}
\expandafter\def\csname ppt@rev.desc.d1.uiv.p1.C.dual-d2.tau-source.tost5.p\endcsname{$+$79.0}
\expandafter\def\csname val@rev.desc.d1.uiv.p1.C.dual-d2.tau-target.ci90.0\endcsname{-0.073}
\expandafter\def\csname cnt@rev.desc.d1.uiv.p1.C.dual-d2.tau-target.ci90.0\endcsname{0}
\expandafter\def\csname pct@rev.desc.d1.uiv.p1.C.dual-d2.tau-target.ci90.0\endcsname{$-$7.3}
\expandafter\def\csname ppt@rev.desc.d1.uiv.p1.C.dual-d2.tau-target.ci90.0\endcsname{$-$7.3}
\expandafter\def\csname val@rev.desc.d1.uiv.p1.C.dual-d2.tau-target.ci90.1\endcsname{0.062}
\expandafter\def\csname cnt@rev.desc.d1.uiv.p1.C.dual-d2.tau-target.ci90.1\endcsname{0}
\expandafter\def\csname pct@rev.desc.d1.uiv.p1.C.dual-d2.tau-target.ci90.1\endcsname{6.2}
\expandafter\def\csname ppt@rev.desc.d1.uiv.p1.C.dual-d2.tau-target.ci90.1\endcsname{$+$6.2}
\expandafter\def\csname val@rev.desc.d1.uiv.p1.C.dual-d2.tau-target.ci95-excludes-0\endcsname{0}
\expandafter\def\csname cnt@rev.desc.d1.uiv.p1.C.dual-d2.tau-target.ci95-excludes-0\endcsname{0}
\expandafter\def\csname pct@rev.desc.d1.uiv.p1.C.dual-d2.tau-target.ci95-excludes-0\endcsname{0.0}
\expandafter\def\csname ppt@rev.desc.d1.uiv.p1.C.dual-d2.tau-target.ci95-excludes-0\endcsname{0.0}
\expandafter\def\csname val@rev.desc.d1.uiv.p1.C.dual-d2.tau-target.ci95.0\endcsname{-0.094}
\expandafter\def\csname cnt@rev.desc.d1.uiv.p1.C.dual-d2.tau-target.ci95.0\endcsname{0}
\expandafter\def\csname pct@rev.desc.d1.uiv.p1.C.dual-d2.tau-target.ci95.0\endcsname{$-$9.4}
\expandafter\def\csname ppt@rev.desc.d1.uiv.p1.C.dual-d2.tau-target.ci95.0\endcsname{$-$9.4}
\expandafter\def\csname val@rev.desc.d1.uiv.p1.C.dual-d2.tau-target.ci95.1\endcsname{0.073}
\expandafter\def\csname cnt@rev.desc.d1.uiv.p1.C.dual-d2.tau-target.ci95.1\endcsname{0}
\expandafter\def\csname pct@rev.desc.d1.uiv.p1.C.dual-d2.tau-target.ci95.1\endcsname{7.3}
\expandafter\def\csname ppt@rev.desc.d1.uiv.p1.C.dual-d2.tau-target.ci95.1\endcsname{$+$7.3}
\expandafter\def\csname val@rev.desc.d1.uiv.p1.C.dual-d2.tau-target.mean\endcsname{0.000}
\expandafter\def\csname cnt@rev.desc.d1.uiv.p1.C.dual-d2.tau-target.mean\endcsname{0}
\expandafter\def\csname pct@rev.desc.d1.uiv.p1.C.dual-d2.tau-target.mean\endcsname{0.0}
\expandafter\def\csname ppt@rev.desc.d1.uiv.p1.C.dual-d2.tau-target.mean\endcsname{0.0}
\expandafter\def\csname val@rev.desc.d1.uiv.p1.C.dual-d2.tau-target.n-apps\endcsname{12}
\expandafter\def\csname cnt@rev.desc.d1.uiv.p1.C.dual-d2.tau-target.n-apps\endcsname{12}
\expandafter\def\csname pct@rev.desc.d1.uiv.p1.C.dual-d2.tau-target.n-apps\endcsname{1200.0}
\expandafter\def\csname ppt@rev.desc.d1.uiv.p1.C.dual-d2.tau-target.n-apps\endcsname{$+$1200.0}
\expandafter\def\csname val@rev.desc.d1.uiv.p1.C.dual-d2.tau-target.tost1.ci90-inside\endcsname{0}
\expandafter\def\csname cnt@rev.desc.d1.uiv.p1.C.dual-d2.tau-target.tost1.ci90-inside\endcsname{0}
\expandafter\def\csname pct@rev.desc.d1.uiv.p1.C.dual-d2.tau-target.tost1.ci90-inside\endcsname{0.0}
\expandafter\def\csname ppt@rev.desc.d1.uiv.p1.C.dual-d2.tau-target.tost1.ci90-inside\endcsname{0.0}
\expandafter\def\csname val@rev.desc.d1.uiv.p1.C.dual-d2.tau-target.tost1.equivalent\endcsname{0}
\expandafter\def\csname cnt@rev.desc.d1.uiv.p1.C.dual-d2.tau-target.tost1.equivalent\endcsname{0}
\expandafter\def\csname pct@rev.desc.d1.uiv.p1.C.dual-d2.tau-target.tost1.equivalent\endcsname{0.0}
\expandafter\def\csname ppt@rev.desc.d1.uiv.p1.C.dual-d2.tau-target.tost1.equivalent\endcsname{0.0}
\expandafter\def\csname val@rev.desc.d1.uiv.p1.C.dual-d2.tau-target.tost1.p\endcsname{0.472}
\expandafter\def\csname cnt@rev.desc.d1.uiv.p1.C.dual-d2.tau-target.tost1.p\endcsname{0}
\expandafter\def\csname pct@rev.desc.d1.uiv.p1.C.dual-d2.tau-target.tost1.p\endcsname{47.2}
\expandafter\def\csname ppt@rev.desc.d1.uiv.p1.C.dual-d2.tau-target.tost1.p\endcsname{$+$47.2}
\expandafter\def\csname val@rev.desc.d1.uiv.p1.C.dual-d2.tau-target.tost5.ci90-inside\endcsname{0}
\expandafter\def\csname cnt@rev.desc.d1.uiv.p1.C.dual-d2.tau-target.tost5.ci90-inside\endcsname{0}
\expandafter\def\csname pct@rev.desc.d1.uiv.p1.C.dual-d2.tau-target.tost5.ci90-inside\endcsname{0.0}
\expandafter\def\csname ppt@rev.desc.d1.uiv.p1.C.dual-d2.tau-target.tost5.ci90-inside\endcsname{0.0}
\expandafter\def\csname val@rev.desc.d1.uiv.p1.C.dual-d2.tau-target.tost5.equivalent\endcsname{0}
\expandafter\def\csname cnt@rev.desc.d1.uiv.p1.C.dual-d2.tau-target.tost5.equivalent\endcsname{0}
\expandafter\def\csname pct@rev.desc.d1.uiv.p1.C.dual-d2.tau-target.tost5.equivalent\endcsname{0.0}
\expandafter\def\csname ppt@rev.desc.d1.uiv.p1.C.dual-d2.tau-target.tost5.equivalent\endcsname{0.0}
\expandafter\def\csname val@rev.desc.d1.uiv.p1.C.dual-d2.tau-target.tost5.p\endcsname{0.139}
\expandafter\def\csname cnt@rev.desc.d1.uiv.p1.C.dual-d2.tau-target.tost5.p\endcsname{0}
\expandafter\def\csname pct@rev.desc.d1.uiv.p1.C.dual-d2.tau-target.tost5.p\endcsname{13.9}
\expandafter\def\csname ppt@rev.desc.d1.uiv.p1.C.dual-d2.tau-target.tost5.p\endcsname{$+$13.9}
\expandafter\def\csname val@rev.desc.d1.uiv.p1.C.dual-d2.zero-estimate\endcsname{1}
\expandafter\def\csname cnt@rev.desc.d1.uiv.p1.C.dual-d2.zero-estimate\endcsname{1}
\expandafter\def\csname pct@rev.desc.d1.uiv.p1.C.dual-d2.zero-estimate\endcsname{100.0}
\expandafter\def\csname ppt@rev.desc.d1.uiv.p1.C.dual-d2.zero-estimate\endcsname{$+$100.0}
\expandafter\def\csname val@rev.desc.d1.uiv.p1.C.grid2-d3.delta-tau.ci90.0\endcsname{0.042}
\expandafter\def\csname cnt@rev.desc.d1.uiv.p1.C.grid2-d3.delta-tau.ci90.0\endcsname{0}
\expandafter\def\csname pct@rev.desc.d1.uiv.p1.C.grid2-d3.delta-tau.ci90.0\endcsname{4.2}
\expandafter\def\csname ppt@rev.desc.d1.uiv.p1.C.grid2-d3.delta-tau.ci90.0\endcsname{$+$4.2}
\expandafter\def\csname val@rev.desc.d1.uiv.p1.C.grid2-d3.delta-tau.ci90.1\endcsname{0.188}
\expandafter\def\csname cnt@rev.desc.d1.uiv.p1.C.grid2-d3.delta-tau.ci90.1\endcsname{0}
\expandafter\def\csname pct@rev.desc.d1.uiv.p1.C.grid2-d3.delta-tau.ci90.1\endcsname{18.8}
\expandafter\def\csname ppt@rev.desc.d1.uiv.p1.C.grid2-d3.delta-tau.ci90.1\endcsname{$+$18.8}
\expandafter\def\csname val@rev.desc.d1.uiv.p1.C.grid2-d3.delta-tau.ci95-excludes-0\endcsname{1}
\expandafter\def\csname cnt@rev.desc.d1.uiv.p1.C.grid2-d3.delta-tau.ci95-excludes-0\endcsname{1}
\expandafter\def\csname pct@rev.desc.d1.uiv.p1.C.grid2-d3.delta-tau.ci95-excludes-0\endcsname{100.0}
\expandafter\def\csname ppt@rev.desc.d1.uiv.p1.C.grid2-d3.delta-tau.ci95-excludes-0\endcsname{$+$100.0}
\expandafter\def\csname val@rev.desc.d1.uiv.p1.C.grid2-d3.delta-tau.ci95.0\endcsname{0.031}
\expandafter\def\csname cnt@rev.desc.d1.uiv.p1.C.grid2-d3.delta-tau.ci95.0\endcsname{0}
\expandafter\def\csname pct@rev.desc.d1.uiv.p1.C.grid2-d3.delta-tau.ci95.0\endcsname{3.1}
\expandafter\def\csname ppt@rev.desc.d1.uiv.p1.C.grid2-d3.delta-tau.ci95.0\endcsname{$+$3.1}
\expandafter\def\csname val@rev.desc.d1.uiv.p1.C.grid2-d3.delta-tau.ci95.1\endcsname{0.198}
\expandafter\def\csname cnt@rev.desc.d1.uiv.p1.C.grid2-d3.delta-tau.ci95.1\endcsname{0}
\expandafter\def\csname pct@rev.desc.d1.uiv.p1.C.grid2-d3.delta-tau.ci95.1\endcsname{19.8}
\expandafter\def\csname ppt@rev.desc.d1.uiv.p1.C.grid2-d3.delta-tau.ci95.1\endcsname{$+$19.8}
\expandafter\def\csname val@rev.desc.d1.uiv.p1.C.grid2-d3.delta-tau.mean\endcsname{0.115}
\expandafter\def\csname cnt@rev.desc.d1.uiv.p1.C.grid2-d3.delta-tau.mean\endcsname{0}
\expandafter\def\csname pct@rev.desc.d1.uiv.p1.C.grid2-d3.delta-tau.mean\endcsname{11.5}
\expandafter\def\csname ppt@rev.desc.d1.uiv.p1.C.grid2-d3.delta-tau.mean\endcsname{$+$11.5}
\expandafter\def\csname val@rev.desc.d1.uiv.p1.C.grid2-d3.delta-tau.n-apps\endcsname{12}
\expandafter\def\csname cnt@rev.desc.d1.uiv.p1.C.grid2-d3.delta-tau.n-apps\endcsname{12}
\expandafter\def\csname pct@rev.desc.d1.uiv.p1.C.grid2-d3.delta-tau.n-apps\endcsname{1200.0}
\expandafter\def\csname ppt@rev.desc.d1.uiv.p1.C.grid2-d3.delta-tau.n-apps\endcsname{$+$1200.0}
\expandafter\def\csname val@rev.desc.d1.uiv.p1.C.grid2-d3.delta-tau.tost1.ci90-inside\endcsname{0}
\expandafter\def\csname cnt@rev.desc.d1.uiv.p1.C.grid2-d3.delta-tau.tost1.ci90-inside\endcsname{0}
\expandafter\def\csname pct@rev.desc.d1.uiv.p1.C.grid2-d3.delta-tau.tost1.ci90-inside\endcsname{0.0}
\expandafter\def\csname ppt@rev.desc.d1.uiv.p1.C.grid2-d3.delta-tau.tost1.ci90-inside\endcsname{0.0}
\expandafter\def\csname val@rev.desc.d1.uiv.p1.C.grid2-d3.delta-tau.tost1.equivalent\endcsname{0}
\expandafter\def\csname cnt@rev.desc.d1.uiv.p1.C.grid2-d3.delta-tau.tost1.equivalent\endcsname{0}
\expandafter\def\csname pct@rev.desc.d1.uiv.p1.C.grid2-d3.delta-tau.tost1.equivalent\endcsname{0.0}
\expandafter\def\csname ppt@rev.desc.d1.uiv.p1.C.grid2-d3.delta-tau.tost1.equivalent\endcsname{0.0}
\expandafter\def\csname val@rev.desc.d1.uiv.p1.C.grid2-d3.delta-tau.tost1.p\endcsname{0.994}
\expandafter\def\csname cnt@rev.desc.d1.uiv.p1.C.grid2-d3.delta-tau.tost1.p\endcsname{1}
\expandafter\def\csname pct@rev.desc.d1.uiv.p1.C.grid2-d3.delta-tau.tost1.p\endcsname{99.4}
\expandafter\def\csname ppt@rev.desc.d1.uiv.p1.C.grid2-d3.delta-tau.tost1.p\endcsname{$+$99.4}
\expandafter\def\csname val@rev.desc.d1.uiv.p1.C.grid2-d3.delta-tau.tost5.ci90-inside\endcsname{0}
\expandafter\def\csname cnt@rev.desc.d1.uiv.p1.C.grid2-d3.delta-tau.tost5.ci90-inside\endcsname{0}
\expandafter\def\csname pct@rev.desc.d1.uiv.p1.C.grid2-d3.delta-tau.tost5.ci90-inside\endcsname{0.0}
\expandafter\def\csname ppt@rev.desc.d1.uiv.p1.C.grid2-d3.delta-tau.tost5.ci90-inside\endcsname{0.0}
\expandafter\def\csname val@rev.desc.d1.uiv.p1.C.grid2-d3.delta-tau.tost5.equivalent\endcsname{0}
\expandafter\def\csname cnt@rev.desc.d1.uiv.p1.C.grid2-d3.delta-tau.tost5.equivalent\endcsname{0}
\expandafter\def\csname pct@rev.desc.d1.uiv.p1.C.grid2-d3.delta-tau.tost5.equivalent\endcsname{0.0}
\expandafter\def\csname ppt@rev.desc.d1.uiv.p1.C.grid2-d3.delta-tau.tost5.equivalent\endcsname{0.0}
\expandafter\def\csname val@rev.desc.d1.uiv.p1.C.grid2-d3.delta-tau.tost5.p\endcsname{0.934}
\expandafter\def\csname cnt@rev.desc.d1.uiv.p1.C.grid2-d3.delta-tau.tost5.p\endcsname{1}
\expandafter\def\csname pct@rev.desc.d1.uiv.p1.C.grid2-d3.delta-tau.tost5.p\endcsname{93.4}
\expandafter\def\csname ppt@rev.desc.d1.uiv.p1.C.grid2-d3.delta-tau.tost5.p\endcsname{$+$93.4}
\expandafter\def\csname val@rev.desc.d1.uiv.p1.C.grid2-d3.material-flip\endcsname{0}
\expandafter\def\csname cnt@rev.desc.d1.uiv.p1.C.grid2-d3.material-flip\endcsname{0}
\expandafter\def\csname pct@rev.desc.d1.uiv.p1.C.grid2-d3.material-flip\endcsname{0.0}
\expandafter\def\csname ppt@rev.desc.d1.uiv.p1.C.grid2-d3.material-flip\endcsname{0.0}
\expandafter\def\csname val@rev.desc.d1.uiv.p1.C.grid2-d3.n-apps\endcsname{12}
\expandafter\def\csname cnt@rev.desc.d1.uiv.p1.C.grid2-d3.n-apps\endcsname{12}
\expandafter\def\csname pct@rev.desc.d1.uiv.p1.C.grid2-d3.n-apps\endcsname{1200.0}
\expandafter\def\csname ppt@rev.desc.d1.uiv.p1.C.grid2-d3.n-apps\endcsname{$+$1200.0}
\expandafter\def\csname val@rev.desc.d1.uiv.p1.C.grid2-d3.n-tasks\endcsname{96}
\expandafter\def\csname cnt@rev.desc.d1.uiv.p1.C.grid2-d3.n-tasks\endcsname{96}
\expandafter\def\csname pct@rev.desc.d1.uiv.p1.C.grid2-d3.n-tasks\endcsname{9600.0}
\expandafter\def\csname ppt@rev.desc.d1.uiv.p1.C.grid2-d3.n-tasks\endcsname{$+$9600.0}
\expandafter\def\csname val@rev.desc.d1.uiv.p1.C.grid2-d3.sign-differs\endcsname{0}
\expandafter\def\csname cnt@rev.desc.d1.uiv.p1.C.grid2-d3.sign-differs\endcsname{0}
\expandafter\def\csname pct@rev.desc.d1.uiv.p1.C.grid2-d3.sign-differs\endcsname{0.0}
\expandafter\def\csname ppt@rev.desc.d1.uiv.p1.C.grid2-d3.sign-differs\endcsname{0.0}
\expandafter\def\csname val@rev.desc.d1.uiv.p1.C.grid2-d3.tau-source.ci90.0\endcsname{-0.208}
\expandafter\def\csname cnt@rev.desc.d1.uiv.p1.C.grid2-d3.tau-source.ci90.0\endcsname{0}
\expandafter\def\csname pct@rev.desc.d1.uiv.p1.C.grid2-d3.tau-source.ci90.0\endcsname{$-$20.8}
\expandafter\def\csname ppt@rev.desc.d1.uiv.p1.C.grid2-d3.tau-source.ci90.0\endcsname{$-$20.8}
\expandafter\def\csname val@rev.desc.d1.uiv.p1.C.grid2-d3.tau-source.ci90.1\endcsname{-0.073}
\expandafter\def\csname cnt@rev.desc.d1.uiv.p1.C.grid2-d3.tau-source.ci90.1\endcsname{0}
\expandafter\def\csname pct@rev.desc.d1.uiv.p1.C.grid2-d3.tau-source.ci90.1\endcsname{$-$7.3}
\expandafter\def\csname ppt@rev.desc.d1.uiv.p1.C.grid2-d3.tau-source.ci90.1\endcsname{$-$7.3}
\expandafter\def\csname val@rev.desc.d1.uiv.p1.C.grid2-d3.tau-source.ci95-excludes-0\endcsname{1}
\expandafter\def\csname cnt@rev.desc.d1.uiv.p1.C.grid2-d3.tau-source.ci95-excludes-0\endcsname{1}
\expandafter\def\csname pct@rev.desc.d1.uiv.p1.C.grid2-d3.tau-source.ci95-excludes-0\endcsname{100.0}
\expandafter\def\csname ppt@rev.desc.d1.uiv.p1.C.grid2-d3.tau-source.ci95-excludes-0\endcsname{$+$100.0}
\expandafter\def\csname val@rev.desc.d1.uiv.p1.C.grid2-d3.tau-source.ci95.0\endcsname{-0.229}
\expandafter\def\csname cnt@rev.desc.d1.uiv.p1.C.grid2-d3.tau-source.ci95.0\endcsname{0}
\expandafter\def\csname pct@rev.desc.d1.uiv.p1.C.grid2-d3.tau-source.ci95.0\endcsname{$-$22.9}
\expandafter\def\csname ppt@rev.desc.d1.uiv.p1.C.grid2-d3.tau-source.ci95.0\endcsname{$-$22.9}
\expandafter\def\csname val@rev.desc.d1.uiv.p1.C.grid2-d3.tau-source.ci95.1\endcsname{-0.062}
\expandafter\def\csname cnt@rev.desc.d1.uiv.p1.C.grid2-d3.tau-source.ci95.1\endcsname{0}
\expandafter\def\csname pct@rev.desc.d1.uiv.p1.C.grid2-d3.tau-source.ci95.1\endcsname{$-$6.2}
\expandafter\def\csname ppt@rev.desc.d1.uiv.p1.C.grid2-d3.tau-source.ci95.1\endcsname{$-$6.2}
\expandafter\def\csname val@rev.desc.d1.uiv.p1.C.grid2-d3.tau-source.mean\endcsname{-0.146}
\expandafter\def\csname cnt@rev.desc.d1.uiv.p1.C.grid2-d3.tau-source.mean\endcsname{0}
\expandafter\def\csname pct@rev.desc.d1.uiv.p1.C.grid2-d3.tau-source.mean\endcsname{$-$14.6}
\expandafter\def\csname ppt@rev.desc.d1.uiv.p1.C.grid2-d3.tau-source.mean\endcsname{$-$14.6}
\expandafter\def\csname val@rev.desc.d1.uiv.p1.C.grid2-d3.tau-source.n-apps\endcsname{12}
\expandafter\def\csname cnt@rev.desc.d1.uiv.p1.C.grid2-d3.tau-source.n-apps\endcsname{12}
\expandafter\def\csname pct@rev.desc.d1.uiv.p1.C.grid2-d3.tau-source.n-apps\endcsname{1200.0}
\expandafter\def\csname ppt@rev.desc.d1.uiv.p1.C.grid2-d3.tau-source.n-apps\endcsname{$+$1200.0}
\expandafter\def\csname val@rev.desc.d1.uiv.p1.C.grid2-d3.tau-source.tost1.ci90-inside\endcsname{0}
\expandafter\def\csname cnt@rev.desc.d1.uiv.p1.C.grid2-d3.tau-source.tost1.ci90-inside\endcsname{0}
\expandafter\def\csname pct@rev.desc.d1.uiv.p1.C.grid2-d3.tau-source.tost1.ci90-inside\endcsname{0.0}
\expandafter\def\csname ppt@rev.desc.d1.uiv.p1.C.grid2-d3.tau-source.tost1.ci90-inside\endcsname{0.0}
\expandafter\def\csname val@rev.desc.d1.uiv.p1.C.grid2-d3.tau-source.tost1.equivalent\endcsname{0}
\expandafter\def\csname cnt@rev.desc.d1.uiv.p1.C.grid2-d3.tau-source.tost1.equivalent\endcsname{0}
\expandafter\def\csname pct@rev.desc.d1.uiv.p1.C.grid2-d3.tau-source.tost1.equivalent\endcsname{0.0}
\expandafter\def\csname ppt@rev.desc.d1.uiv.p1.C.grid2-d3.tau-source.tost1.equivalent\endcsname{0.0}
\expandafter\def\csname val@rev.desc.d1.uiv.p1.C.grid2-d3.tau-source.tost1.p\endcsname{0.999}
\expandafter\def\csname cnt@rev.desc.d1.uiv.p1.C.grid2-d3.tau-source.tost1.p\endcsname{1}
\expandafter\def\csname pct@rev.desc.d1.uiv.p1.C.grid2-d3.tau-source.tost1.p\endcsname{99.9}
\expandafter\def\csname ppt@rev.desc.d1.uiv.p1.C.grid2-d3.tau-source.tost1.p\endcsname{$+$99.9}
\expandafter\def\csname val@rev.desc.d1.uiv.p1.C.grid2-d3.tau-source.tost5.ci90-inside\endcsname{0}
\expandafter\def\csname cnt@rev.desc.d1.uiv.p1.C.grid2-d3.tau-source.tost5.ci90-inside\endcsname{0}
\expandafter\def\csname pct@rev.desc.d1.uiv.p1.C.grid2-d3.tau-source.tost5.ci90-inside\endcsname{0.0}
\expandafter\def\csname ppt@rev.desc.d1.uiv.p1.C.grid2-d3.tau-source.tost5.ci90-inside\endcsname{0.0}
\expandafter\def\csname val@rev.desc.d1.uiv.p1.C.grid2-d3.tau-source.tost5.equivalent\endcsname{0}
\expandafter\def\csname cnt@rev.desc.d1.uiv.p1.C.grid2-d3.tau-source.tost5.equivalent\endcsname{0}
\expandafter\def\csname pct@rev.desc.d1.uiv.p1.C.grid2-d3.tau-source.tost5.equivalent\endcsname{0.0}
\expandafter\def\csname ppt@rev.desc.d1.uiv.p1.C.grid2-d3.tau-source.tost5.equivalent\endcsname{0.0}
\expandafter\def\csname val@rev.desc.d1.uiv.p1.C.grid2-d3.tau-source.tost5.p\endcsname{0.992}
\expandafter\def\csname cnt@rev.desc.d1.uiv.p1.C.grid2-d3.tau-source.tost5.p\endcsname{1}
\expandafter\def\csname pct@rev.desc.d1.uiv.p1.C.grid2-d3.tau-source.tost5.p\endcsname{99.2}
\expandafter\def\csname ppt@rev.desc.d1.uiv.p1.C.grid2-d3.tau-source.tost5.p\endcsname{$+$99.2}
\expandafter\def\csname val@rev.desc.d1.uiv.p1.C.grid2-d3.tau-target.ci90.0\endcsname{-0.094}
\expandafter\def\csname cnt@rev.desc.d1.uiv.p1.C.grid2-d3.tau-target.ci90.0\endcsname{0}
\expandafter\def\csname pct@rev.desc.d1.uiv.p1.C.grid2-d3.tau-target.ci90.0\endcsname{$-$9.4}
\expandafter\def\csname ppt@rev.desc.d1.uiv.p1.C.grid2-d3.tau-target.ci90.0\endcsname{$-$9.4}
\expandafter\def\csname val@rev.desc.d1.uiv.p1.C.grid2-d3.tau-target.ci90.1\endcsname{0.031}
\expandafter\def\csname cnt@rev.desc.d1.uiv.p1.C.grid2-d3.tau-target.ci90.1\endcsname{0}
\expandafter\def\csname pct@rev.desc.d1.uiv.p1.C.grid2-d3.tau-target.ci90.1\endcsname{3.1}
\expandafter\def\csname ppt@rev.desc.d1.uiv.p1.C.grid2-d3.tau-target.ci90.1\endcsname{$+$3.1}
\expandafter\def\csname val@rev.desc.d1.uiv.p1.C.grid2-d3.tau-target.ci95-excludes-0\endcsname{0}
\expandafter\def\csname cnt@rev.desc.d1.uiv.p1.C.grid2-d3.tau-target.ci95-excludes-0\endcsname{0}
\expandafter\def\csname pct@rev.desc.d1.uiv.p1.C.grid2-d3.tau-target.ci95-excludes-0\endcsname{0.0}
\expandafter\def\csname ppt@rev.desc.d1.uiv.p1.C.grid2-d3.tau-target.ci95-excludes-0\endcsname{0.0}
\expandafter\def\csname val@rev.desc.d1.uiv.p1.C.grid2-d3.tau-target.ci95.0\endcsname{-0.104}
\expandafter\def\csname cnt@rev.desc.d1.uiv.p1.C.grid2-d3.tau-target.ci95.0\endcsname{0}
\expandafter\def\csname pct@rev.desc.d1.uiv.p1.C.grid2-d3.tau-target.ci95.0\endcsname{$-$10.4}
\expandafter\def\csname ppt@rev.desc.d1.uiv.p1.C.grid2-d3.tau-target.ci95.0\endcsname{$-$10.4}
\expandafter\def\csname val@rev.desc.d1.uiv.p1.C.grid2-d3.tau-target.ci95.1\endcsname{0.042}
\expandafter\def\csname cnt@rev.desc.d1.uiv.p1.C.grid2-d3.tau-target.ci95.1\endcsname{0}
\expandafter\def\csname pct@rev.desc.d1.uiv.p1.C.grid2-d3.tau-target.ci95.1\endcsname{4.2}
\expandafter\def\csname ppt@rev.desc.d1.uiv.p1.C.grid2-d3.tau-target.ci95.1\endcsname{$+$4.2}
\expandafter\def\csname val@rev.desc.d1.uiv.p1.C.grid2-d3.tau-target.mean\endcsname{-0.031}
\expandafter\def\csname cnt@rev.desc.d1.uiv.p1.C.grid2-d3.tau-target.mean\endcsname{0}
\expandafter\def\csname pct@rev.desc.d1.uiv.p1.C.grid2-d3.tau-target.mean\endcsname{$-$3.1}
\expandafter\def\csname ppt@rev.desc.d1.uiv.p1.C.grid2-d3.tau-target.mean\endcsname{$-$3.1}
\expandafter\def\csname val@rev.desc.d1.uiv.p1.C.grid2-d3.tau-target.n-apps\endcsname{12}
\expandafter\def\csname cnt@rev.desc.d1.uiv.p1.C.grid2-d3.tau-target.n-apps\endcsname{12}
\expandafter\def\csname pct@rev.desc.d1.uiv.p1.C.grid2-d3.tau-target.n-apps\endcsname{1200.0}
\expandafter\def\csname ppt@rev.desc.d1.uiv.p1.C.grid2-d3.tau-target.n-apps\endcsname{$+$1200.0}
\expandafter\def\csname val@rev.desc.d1.uiv.p1.C.grid2-d3.tau-target.tost1.ci90-inside\endcsname{0}
\expandafter\def\csname cnt@rev.desc.d1.uiv.p1.C.grid2-d3.tau-target.tost1.ci90-inside\endcsname{0}
\expandafter\def\csname pct@rev.desc.d1.uiv.p1.C.grid2-d3.tau-target.tost1.ci90-inside\endcsname{0.0}
\expandafter\def\csname ppt@rev.desc.d1.uiv.p1.C.grid2-d3.tau-target.tost1.ci90-inside\endcsname{0.0}
\expandafter\def\csname val@rev.desc.d1.uiv.p1.C.grid2-d3.tau-target.tost1.equivalent\endcsname{0}
\expandafter\def\csname cnt@rev.desc.d1.uiv.p1.C.grid2-d3.tau-target.tost1.equivalent\endcsname{0}
\expandafter\def\csname pct@rev.desc.d1.uiv.p1.C.grid2-d3.tau-target.tost1.equivalent\endcsname{0.0}
\expandafter\def\csname ppt@rev.desc.d1.uiv.p1.C.grid2-d3.tau-target.tost1.equivalent\endcsname{0.0}
\expandafter\def\csname val@rev.desc.d1.uiv.p1.C.grid2-d3.tau-target.tost1.p\endcsname{0.757}
\expandafter\def\csname cnt@rev.desc.d1.uiv.p1.C.grid2-d3.tau-target.tost1.p\endcsname{1}
\expandafter\def\csname pct@rev.desc.d1.uiv.p1.C.grid2-d3.tau-target.tost1.p\endcsname{75.7}
\expandafter\def\csname ppt@rev.desc.d1.uiv.p1.C.grid2-d3.tau-target.tost1.p\endcsname{$+$75.7}
\expandafter\def\csname val@rev.desc.d1.uiv.p1.C.grid2-d3.tau-target.tost5.ci90-inside\endcsname{0}
\expandafter\def\csname cnt@rev.desc.d1.uiv.p1.C.grid2-d3.tau-target.tost5.ci90-inside\endcsname{0}
\expandafter\def\csname pct@rev.desc.d1.uiv.p1.C.grid2-d3.tau-target.tost5.ci90-inside\endcsname{0.0}
\expandafter\def\csname ppt@rev.desc.d1.uiv.p1.C.grid2-d3.tau-target.tost5.ci90-inside\endcsname{0.0}
\expandafter\def\csname val@rev.desc.d1.uiv.p1.C.grid2-d3.tau-target.tost5.equivalent\endcsname{0}
\expandafter\def\csname cnt@rev.desc.d1.uiv.p1.C.grid2-d3.tau-target.tost5.equivalent\endcsname{0}
\expandafter\def\csname pct@rev.desc.d1.uiv.p1.C.grid2-d3.tau-target.tost5.equivalent\endcsname{0.0}
\expandafter\def\csname ppt@rev.desc.d1.uiv.p1.C.grid2-d3.tau-target.tost5.equivalent\endcsname{0.0}
\expandafter\def\csname val@rev.desc.d1.uiv.p1.C.grid2-d3.tau-target.tost5.p\endcsname{0.334}
\expandafter\def\csname cnt@rev.desc.d1.uiv.p1.C.grid2-d3.tau-target.tost5.p\endcsname{0}
\expandafter\def\csname pct@rev.desc.d1.uiv.p1.C.grid2-d3.tau-target.tost5.p\endcsname{33.4}
\expandafter\def\csname ppt@rev.desc.d1.uiv.p1.C.grid2-d3.tau-target.tost5.p\endcsname{$+$33.4}
\expandafter\def\csname val@rev.desc.d1.uiv.p1.C.grid2-d3.zero-estimate\endcsname{0}
\expandafter\def\csname cnt@rev.desc.d1.uiv.p1.C.grid2-d3.zero-estimate\endcsname{0}
\expandafter\def\csname pct@rev.desc.d1.uiv.p1.C.grid2-d3.zero-estimate\endcsname{0.0}
\expandafter\def\csname ppt@rev.desc.d1.uiv.p1.C.grid2-d3.zero-estimate\endcsname{0.0}
\expandafter\def\csname val@rev.desc.d1.uiv.p1.C.zoomclick-d3.delta-tau.ci90.0\endcsname{-0.073}
\expandafter\def\csname cnt@rev.desc.d1.uiv.p1.C.zoomclick-d3.delta-tau.ci90.0\endcsname{0}
\expandafter\def\csname pct@rev.desc.d1.uiv.p1.C.zoomclick-d3.delta-tau.ci90.0\endcsname{$-$7.3}
\expandafter\def\csname ppt@rev.desc.d1.uiv.p1.C.zoomclick-d3.delta-tau.ci90.0\endcsname{$-$7.3}
\expandafter\def\csname val@rev.desc.d1.uiv.p1.C.zoomclick-d3.delta-tau.ci90.1\endcsname{0.094}
\expandafter\def\csname cnt@rev.desc.d1.uiv.p1.C.zoomclick-d3.delta-tau.ci90.1\endcsname{0}
\expandafter\def\csname pct@rev.desc.d1.uiv.p1.C.zoomclick-d3.delta-tau.ci90.1\endcsname{9.4}
\expandafter\def\csname ppt@rev.desc.d1.uiv.p1.C.zoomclick-d3.delta-tau.ci90.1\endcsname{$+$9.4}
\expandafter\def\csname val@rev.desc.d1.uiv.p1.C.zoomclick-d3.delta-tau.ci95-excludes-0\endcsname{0}
\expandafter\def\csname cnt@rev.desc.d1.uiv.p1.C.zoomclick-d3.delta-tau.ci95-excludes-0\endcsname{0}
\expandafter\def\csname pct@rev.desc.d1.uiv.p1.C.zoomclick-d3.delta-tau.ci95-excludes-0\endcsname{0.0}
\expandafter\def\csname ppt@rev.desc.d1.uiv.p1.C.zoomclick-d3.delta-tau.ci95-excludes-0\endcsname{0.0}
\expandafter\def\csname val@rev.desc.d1.uiv.p1.C.zoomclick-d3.delta-tau.ci95.0\endcsname{-0.083}
\expandafter\def\csname cnt@rev.desc.d1.uiv.p1.C.zoomclick-d3.delta-tau.ci95.0\endcsname{0}
\expandafter\def\csname pct@rev.desc.d1.uiv.p1.C.zoomclick-d3.delta-tau.ci95.0\endcsname{$-$8.3}
\expandafter\def\csname ppt@rev.desc.d1.uiv.p1.C.zoomclick-d3.delta-tau.ci95.0\endcsname{$-$8.3}
\expandafter\def\csname val@rev.desc.d1.uiv.p1.C.zoomclick-d3.delta-tau.ci95.1\endcsname{0.104}
\expandafter\def\csname cnt@rev.desc.d1.uiv.p1.C.zoomclick-d3.delta-tau.ci95.1\endcsname{0}
\expandafter\def\csname pct@rev.desc.d1.uiv.p1.C.zoomclick-d3.delta-tau.ci95.1\endcsname{10.4}
\expandafter\def\csname ppt@rev.desc.d1.uiv.p1.C.zoomclick-d3.delta-tau.ci95.1\endcsname{$+$10.4}
\expandafter\def\csname val@rev.desc.d1.uiv.p1.C.zoomclick-d3.delta-tau.mean\endcsname{0.010}
\expandafter\def\csname cnt@rev.desc.d1.uiv.p1.C.zoomclick-d3.delta-tau.mean\endcsname{0}
\expandafter\def\csname pct@rev.desc.d1.uiv.p1.C.zoomclick-d3.delta-tau.mean\endcsname{1.0}
\expandafter\def\csname ppt@rev.desc.d1.uiv.p1.C.zoomclick-d3.delta-tau.mean\endcsname{$+$1.0}
\expandafter\def\csname val@rev.desc.d1.uiv.p1.C.zoomclick-d3.delta-tau.n-apps\endcsname{12}
\expandafter\def\csname cnt@rev.desc.d1.uiv.p1.C.zoomclick-d3.delta-tau.n-apps\endcsname{12}
\expandafter\def\csname pct@rev.desc.d1.uiv.p1.C.zoomclick-d3.delta-tau.n-apps\endcsname{1200.0}
\expandafter\def\csname ppt@rev.desc.d1.uiv.p1.C.zoomclick-d3.delta-tau.n-apps\endcsname{$+$1200.0}
\expandafter\def\csname val@rev.desc.d1.uiv.p1.C.zoomclick-d3.delta-tau.tost1.ci90-inside\endcsname{0}
\expandafter\def\csname cnt@rev.desc.d1.uiv.p1.C.zoomclick-d3.delta-tau.tost1.ci90-inside\endcsname{0}
\expandafter\def\csname pct@rev.desc.d1.uiv.p1.C.zoomclick-d3.delta-tau.tost1.ci90-inside\endcsname{0.0}
\expandafter\def\csname ppt@rev.desc.d1.uiv.p1.C.zoomclick-d3.delta-tau.tost1.ci90-inside\endcsname{0.0}
\expandafter\def\csname val@rev.desc.d1.uiv.p1.C.zoomclick-d3.delta-tau.tost1.equivalent\endcsname{0}
\expandafter\def\csname cnt@rev.desc.d1.uiv.p1.C.zoomclick-d3.delta-tau.tost1.equivalent\endcsname{0}
\expandafter\def\csname pct@rev.desc.d1.uiv.p1.C.zoomclick-d3.delta-tau.tost1.equivalent\endcsname{0.0}
\expandafter\def\csname ppt@rev.desc.d1.uiv.p1.C.zoomclick-d3.delta-tau.tost1.equivalent\endcsname{0.0}
\expandafter\def\csname val@rev.desc.d1.uiv.p1.C.zoomclick-d3.delta-tau.tost1.p\endcsname{0.525}
\expandafter\def\csname cnt@rev.desc.d1.uiv.p1.C.zoomclick-d3.delta-tau.tost1.p\endcsname{1}
\expandafter\def\csname pct@rev.desc.d1.uiv.p1.C.zoomclick-d3.delta-tau.tost1.p\endcsname{52.5}
\expandafter\def\csname ppt@rev.desc.d1.uiv.p1.C.zoomclick-d3.delta-tau.tost1.p\endcsname{$+$52.5}
\expandafter\def\csname val@rev.desc.d1.uiv.p1.C.zoomclick-d3.delta-tau.tost5.ci90-inside\endcsname{0}
\expandafter\def\csname cnt@rev.desc.d1.uiv.p1.C.zoomclick-d3.delta-tau.tost5.ci90-inside\endcsname{0}
\expandafter\def\csname pct@rev.desc.d1.uiv.p1.C.zoomclick-d3.delta-tau.tost5.ci90-inside\endcsname{0.0}
\expandafter\def\csname ppt@rev.desc.d1.uiv.p1.C.zoomclick-d3.delta-tau.tost5.ci90-inside\endcsname{0.0}
\expandafter\def\csname val@rev.desc.d1.uiv.p1.C.zoomclick-d3.delta-tau.tost5.equivalent\endcsname{0}
\expandafter\def\csname cnt@rev.desc.d1.uiv.p1.C.zoomclick-d3.delta-tau.tost5.equivalent\endcsname{0}
\expandafter\def\csname pct@rev.desc.d1.uiv.p1.C.zoomclick-d3.delta-tau.tost5.equivalent\endcsname{0.0}
\expandafter\def\csname ppt@rev.desc.d1.uiv.p1.C.zoomclick-d3.delta-tau.tost5.equivalent\endcsname{0.0}
\expandafter\def\csname val@rev.desc.d1.uiv.p1.C.zoomclick-d3.delta-tau.tost5.p\endcsname{0.224}
\expandafter\def\csname cnt@rev.desc.d1.uiv.p1.C.zoomclick-d3.delta-tau.tost5.p\endcsname{0}
\expandafter\def\csname pct@rev.desc.d1.uiv.p1.C.zoomclick-d3.delta-tau.tost5.p\endcsname{22.4}
\expandafter\def\csname ppt@rev.desc.d1.uiv.p1.C.zoomclick-d3.delta-tau.tost5.p\endcsname{$+$22.4}
\expandafter\def\csname val@rev.desc.d1.uiv.p1.C.zoomclick-d3.material-flip\endcsname{0}
\expandafter\def\csname cnt@rev.desc.d1.uiv.p1.C.zoomclick-d3.material-flip\endcsname{0}
\expandafter\def\csname pct@rev.desc.d1.uiv.p1.C.zoomclick-d3.material-flip\endcsname{0.0}
\expandafter\def\csname ppt@rev.desc.d1.uiv.p1.C.zoomclick-d3.material-flip\endcsname{0.0}
\expandafter\def\csname val@rev.desc.d1.uiv.p1.C.zoomclick-d3.n-apps\endcsname{12}
\expandafter\def\csname cnt@rev.desc.d1.uiv.p1.C.zoomclick-d3.n-apps\endcsname{12}
\expandafter\def\csname pct@rev.desc.d1.uiv.p1.C.zoomclick-d3.n-apps\endcsname{1200.0}
\expandafter\def\csname ppt@rev.desc.d1.uiv.p1.C.zoomclick-d3.n-apps\endcsname{$+$1200.0}
\expandafter\def\csname val@rev.desc.d1.uiv.p1.C.zoomclick-d3.n-tasks\endcsname{96}
\expandafter\def\csname cnt@rev.desc.d1.uiv.p1.C.zoomclick-d3.n-tasks\endcsname{96}
\expandafter\def\csname pct@rev.desc.d1.uiv.p1.C.zoomclick-d3.n-tasks\endcsname{9600.0}
\expandafter\def\csname ppt@rev.desc.d1.uiv.p1.C.zoomclick-d3.n-tasks\endcsname{$+$9600.0}
\expandafter\def\csname val@rev.desc.d1.uiv.p1.C.zoomclick-d3.sign-differs\endcsname{0}
\expandafter\def\csname cnt@rev.desc.d1.uiv.p1.C.zoomclick-d3.sign-differs\endcsname{0}
\expandafter\def\csname pct@rev.desc.d1.uiv.p1.C.zoomclick-d3.sign-differs\endcsname{0.0}
\expandafter\def\csname ppt@rev.desc.d1.uiv.p1.C.zoomclick-d3.sign-differs\endcsname{0.0}
\expandafter\def\csname val@rev.desc.d1.uiv.p1.C.zoomclick-d3.tau-source.ci90.0\endcsname{-0.073}
\expandafter\def\csname cnt@rev.desc.d1.uiv.p1.C.zoomclick-d3.tau-source.ci90.0\endcsname{0}
\expandafter\def\csname pct@rev.desc.d1.uiv.p1.C.zoomclick-d3.tau-source.ci90.0\endcsname{$-$7.3}
\expandafter\def\csname ppt@rev.desc.d1.uiv.p1.C.zoomclick-d3.tau-source.ci90.0\endcsname{$-$7.3}
\expandafter\def\csname val@rev.desc.d1.uiv.p1.C.zoomclick-d3.tau-source.ci90.1\endcsname{0.010}
\expandafter\def\csname cnt@rev.desc.d1.uiv.p1.C.zoomclick-d3.tau-source.ci90.1\endcsname{0}
\expandafter\def\csname pct@rev.desc.d1.uiv.p1.C.zoomclick-d3.tau-source.ci90.1\endcsname{1.0}
\expandafter\def\csname ppt@rev.desc.d1.uiv.p1.C.zoomclick-d3.tau-source.ci90.1\endcsname{$+$1.0}
\expandafter\def\csname val@rev.desc.d1.uiv.p1.C.zoomclick-d3.tau-source.ci95-excludes-0\endcsname{0}
\expandafter\def\csname cnt@rev.desc.d1.uiv.p1.C.zoomclick-d3.tau-source.ci95-excludes-0\endcsname{0}
\expandafter\def\csname pct@rev.desc.d1.uiv.p1.C.zoomclick-d3.tau-source.ci95-excludes-0\endcsname{0.0}
\expandafter\def\csname ppt@rev.desc.d1.uiv.p1.C.zoomclick-d3.tau-source.ci95-excludes-0\endcsname{0.0}
\expandafter\def\csname val@rev.desc.d1.uiv.p1.C.zoomclick-d3.tau-source.ci95.0\endcsname{-0.083}
\expandafter\def\csname cnt@rev.desc.d1.uiv.p1.C.zoomclick-d3.tau-source.ci95.0\endcsname{0}
\expandafter\def\csname pct@rev.desc.d1.uiv.p1.C.zoomclick-d3.tau-source.ci95.0\endcsname{$-$8.3}
\expandafter\def\csname ppt@rev.desc.d1.uiv.p1.C.zoomclick-d3.tau-source.ci95.0\endcsname{$-$8.3}
\expandafter\def\csname val@rev.desc.d1.uiv.p1.C.zoomclick-d3.tau-source.ci95.1\endcsname{0.021}
\expandafter\def\csname cnt@rev.desc.d1.uiv.p1.C.zoomclick-d3.tau-source.ci95.1\endcsname{0}
\expandafter\def\csname pct@rev.desc.d1.uiv.p1.C.zoomclick-d3.tau-source.ci95.1\endcsname{2.1}
\expandafter\def\csname ppt@rev.desc.d1.uiv.p1.C.zoomclick-d3.tau-source.ci95.1\endcsname{$+$2.1}
\expandafter\def\csname val@rev.desc.d1.uiv.p1.C.zoomclick-d3.tau-source.mean\endcsname{-0.031}
\expandafter\def\csname cnt@rev.desc.d1.uiv.p1.C.zoomclick-d3.tau-source.mean\endcsname{0}
\expandafter\def\csname pct@rev.desc.d1.uiv.p1.C.zoomclick-d3.tau-source.mean\endcsname{$-$3.1}
\expandafter\def\csname ppt@rev.desc.d1.uiv.p1.C.zoomclick-d3.tau-source.mean\endcsname{$-$3.1}
\expandafter\def\csname val@rev.desc.d1.uiv.p1.C.zoomclick-d3.tau-source.n-apps\endcsname{12}
\expandafter\def\csname cnt@rev.desc.d1.uiv.p1.C.zoomclick-d3.tau-source.n-apps\endcsname{12}
\expandafter\def\csname pct@rev.desc.d1.uiv.p1.C.zoomclick-d3.tau-source.n-apps\endcsname{1200.0}
\expandafter\def\csname ppt@rev.desc.d1.uiv.p1.C.zoomclick-d3.tau-source.n-apps\endcsname{$+$1200.0}
\expandafter\def\csname val@rev.desc.d1.uiv.p1.C.zoomclick-d3.tau-source.tost1.ci90-inside\endcsname{0}
\expandafter\def\csname cnt@rev.desc.d1.uiv.p1.C.zoomclick-d3.tau-source.tost1.ci90-inside\endcsname{0}
\expandafter\def\csname pct@rev.desc.d1.uiv.p1.C.zoomclick-d3.tau-source.tost1.ci90-inside\endcsname{0.0}
\expandafter\def\csname ppt@rev.desc.d1.uiv.p1.C.zoomclick-d3.tau-source.tost1.ci90-inside\endcsname{0.0}
\expandafter\def\csname val@rev.desc.d1.uiv.p1.C.zoomclick-d3.tau-source.tost1.equivalent\endcsname{0}
\expandafter\def\csname cnt@rev.desc.d1.uiv.p1.C.zoomclick-d3.tau-source.tost1.equivalent\endcsname{0}
\expandafter\def\csname pct@rev.desc.d1.uiv.p1.C.zoomclick-d3.tau-source.tost1.equivalent\endcsname{0.0}
\expandafter\def\csname ppt@rev.desc.d1.uiv.p1.C.zoomclick-d3.tau-source.tost1.equivalent\endcsname{0.0}
\expandafter\def\csname val@rev.desc.d1.uiv.p1.C.zoomclick-d3.tau-source.tost1.p\endcsname{0.840}
\expandafter\def\csname cnt@rev.desc.d1.uiv.p1.C.zoomclick-d3.tau-source.tost1.p\endcsname{1}
\expandafter\def\csname pct@rev.desc.d1.uiv.p1.C.zoomclick-d3.tau-source.tost1.p\endcsname{84.0}
\expandafter\def\csname ppt@rev.desc.d1.uiv.p1.C.zoomclick-d3.tau-source.tost1.p\endcsname{$+$84.0}
\expandafter\def\csname val@rev.desc.d1.uiv.p1.C.zoomclick-d3.tau-source.tost5.ci90-inside\endcsname{0}
\expandafter\def\csname cnt@rev.desc.d1.uiv.p1.C.zoomclick-d3.tau-source.tost5.ci90-inside\endcsname{0}
\expandafter\def\csname pct@rev.desc.d1.uiv.p1.C.zoomclick-d3.tau-source.tost5.ci90-inside\endcsname{0.0}
\expandafter\def\csname ppt@rev.desc.d1.uiv.p1.C.zoomclick-d3.tau-source.tost5.ci90-inside\endcsname{0.0}
\expandafter\def\csname val@rev.desc.d1.uiv.p1.C.zoomclick-d3.tau-source.tost5.equivalent\endcsname{0}
\expandafter\def\csname cnt@rev.desc.d1.uiv.p1.C.zoomclick-d3.tau-source.tost5.equivalent\endcsname{0}
\expandafter\def\csname pct@rev.desc.d1.uiv.p1.C.zoomclick-d3.tau-source.tost5.equivalent\endcsname{0.0}
\expandafter\def\csname ppt@rev.desc.d1.uiv.p1.C.zoomclick-d3.tau-source.tost5.equivalent\endcsname{0.0}
\expandafter\def\csname val@rev.desc.d1.uiv.p1.C.zoomclick-d3.tau-source.tost5.p\endcsname{0.264}
\expandafter\def\csname cnt@rev.desc.d1.uiv.p1.C.zoomclick-d3.tau-source.tost5.p\endcsname{0}
\expandafter\def\csname pct@rev.desc.d1.uiv.p1.C.zoomclick-d3.tau-source.tost5.p\endcsname{26.4}
\expandafter\def\csname ppt@rev.desc.d1.uiv.p1.C.zoomclick-d3.tau-source.tost5.p\endcsname{$+$26.4}
\expandafter\def\csname val@rev.desc.d1.uiv.p1.C.zoomclick-d3.tau-target.ci90.0\endcsname{-0.094}
\expandafter\def\csname cnt@rev.desc.d1.uiv.p1.C.zoomclick-d3.tau-target.ci90.0\endcsname{0}
\expandafter\def\csname pct@rev.desc.d1.uiv.p1.C.zoomclick-d3.tau-target.ci90.0\endcsname{$-$9.4}
\expandafter\def\csname ppt@rev.desc.d1.uiv.p1.C.zoomclick-d3.tau-target.ci90.0\endcsname{$-$9.4}
\expandafter\def\csname val@rev.desc.d1.uiv.p1.C.zoomclick-d3.tau-target.ci90.1\endcsname{0.042}
\expandafter\def\csname cnt@rev.desc.d1.uiv.p1.C.zoomclick-d3.tau-target.ci90.1\endcsname{0}
\expandafter\def\csname pct@rev.desc.d1.uiv.p1.C.zoomclick-d3.tau-target.ci90.1\endcsname{4.2}
\expandafter\def\csname ppt@rev.desc.d1.uiv.p1.C.zoomclick-d3.tau-target.ci90.1\endcsname{$+$4.2}
\expandafter\def\csname val@rev.desc.d1.uiv.p1.C.zoomclick-d3.tau-target.ci95-excludes-0\endcsname{0}
\expandafter\def\csname cnt@rev.desc.d1.uiv.p1.C.zoomclick-d3.tau-target.ci95-excludes-0\endcsname{0}
\expandafter\def\csname pct@rev.desc.d1.uiv.p1.C.zoomclick-d3.tau-target.ci95-excludes-0\endcsname{0.0}
\expandafter\def\csname ppt@rev.desc.d1.uiv.p1.C.zoomclick-d3.tau-target.ci95-excludes-0\endcsname{0.0}
\expandafter\def\csname val@rev.desc.d1.uiv.p1.C.zoomclick-d3.tau-target.ci95.0\endcsname{-0.104}
\expandafter\def\csname cnt@rev.desc.d1.uiv.p1.C.zoomclick-d3.tau-target.ci95.0\endcsname{0}
\expandafter\def\csname pct@rev.desc.d1.uiv.p1.C.zoomclick-d3.tau-target.ci95.0\endcsname{$-$10.4}
\expandafter\def\csname ppt@rev.desc.d1.uiv.p1.C.zoomclick-d3.tau-target.ci95.0\endcsname{$-$10.4}
\expandafter\def\csname val@rev.desc.d1.uiv.p1.C.zoomclick-d3.tau-target.ci95.1\endcsname{0.052}
\expandafter\def\csname cnt@rev.desc.d1.uiv.p1.C.zoomclick-d3.tau-target.ci95.1\endcsname{0}
\expandafter\def\csname pct@rev.desc.d1.uiv.p1.C.zoomclick-d3.tau-target.ci95.1\endcsname{5.2}
\expandafter\def\csname ppt@rev.desc.d1.uiv.p1.C.zoomclick-d3.tau-target.ci95.1\endcsname{$+$5.2}
\expandafter\def\csname val@rev.desc.d1.uiv.p1.C.zoomclick-d3.tau-target.mean\endcsname{-0.021}
\expandafter\def\csname cnt@rev.desc.d1.uiv.p1.C.zoomclick-d3.tau-target.mean\endcsname{0}
\expandafter\def\csname pct@rev.desc.d1.uiv.p1.C.zoomclick-d3.tau-target.mean\endcsname{$-$2.1}
\expandafter\def\csname ppt@rev.desc.d1.uiv.p1.C.zoomclick-d3.tau-target.mean\endcsname{$-$2.1}
\expandafter\def\csname val@rev.desc.d1.uiv.p1.C.zoomclick-d3.tau-target.n-apps\endcsname{12}
\expandafter\def\csname cnt@rev.desc.d1.uiv.p1.C.zoomclick-d3.tau-target.n-apps\endcsname{12}
\expandafter\def\csname pct@rev.desc.d1.uiv.p1.C.zoomclick-d3.tau-target.n-apps\endcsname{1200.0}
\expandafter\def\csname ppt@rev.desc.d1.uiv.p1.C.zoomclick-d3.tau-target.n-apps\endcsname{$+$1200.0}
\expandafter\def\csname val@rev.desc.d1.uiv.p1.C.zoomclick-d3.tau-target.tost1.ci90-inside\endcsname{0}
\expandafter\def\csname cnt@rev.desc.d1.uiv.p1.C.zoomclick-d3.tau-target.tost1.ci90-inside\endcsname{0}
\expandafter\def\csname pct@rev.desc.d1.uiv.p1.C.zoomclick-d3.tau-target.tost1.ci90-inside\endcsname{0.0}
\expandafter\def\csname ppt@rev.desc.d1.uiv.p1.C.zoomclick-d3.tau-target.tost1.ci90-inside\endcsname{0.0}
\expandafter\def\csname val@rev.desc.d1.uiv.p1.C.zoomclick-d3.tau-target.tost1.equivalent\endcsname{0}
\expandafter\def\csname cnt@rev.desc.d1.uiv.p1.C.zoomclick-d3.tau-target.tost1.equivalent\endcsname{0}
\expandafter\def\csname pct@rev.desc.d1.uiv.p1.C.zoomclick-d3.tau-target.tost1.equivalent\endcsname{0.0}
\expandafter\def\csname ppt@rev.desc.d1.uiv.p1.C.zoomclick-d3.tau-target.tost1.equivalent\endcsname{0.0}
\expandafter\def\csname val@rev.desc.d1.uiv.p1.C.zoomclick-d3.tau-target.tost1.p\endcsname{0.634}
\expandafter\def\csname cnt@rev.desc.d1.uiv.p1.C.zoomclick-d3.tau-target.tost1.p\endcsname{1}
\expandafter\def\csname pct@rev.desc.d1.uiv.p1.C.zoomclick-d3.tau-target.tost1.p\endcsname{63.4}
\expandafter\def\csname ppt@rev.desc.d1.uiv.p1.C.zoomclick-d3.tau-target.tost1.p\endcsname{$+$63.4}
\expandafter\def\csname val@rev.desc.d1.uiv.p1.C.zoomclick-d3.tau-target.tost5.ci90-inside\endcsname{0}
\expandafter\def\csname cnt@rev.desc.d1.uiv.p1.C.zoomclick-d3.tau-target.tost5.ci90-inside\endcsname{0}
\expandafter\def\csname pct@rev.desc.d1.uiv.p1.C.zoomclick-d3.tau-target.tost5.ci90-inside\endcsname{0.0}
\expandafter\def\csname ppt@rev.desc.d1.uiv.p1.C.zoomclick-d3.tau-target.tost5.ci90-inside\endcsname{0.0}
\expandafter\def\csname val@rev.desc.d1.uiv.p1.C.zoomclick-d3.tau-target.tost5.equivalent\endcsname{0}
\expandafter\def\csname cnt@rev.desc.d1.uiv.p1.C.zoomclick-d3.tau-target.tost5.equivalent\endcsname{0}
\expandafter\def\csname pct@rev.desc.d1.uiv.p1.C.zoomclick-d3.tau-target.tost5.equivalent\endcsname{0.0}
\expandafter\def\csname ppt@rev.desc.d1.uiv.p1.C.zoomclick-d3.tau-target.tost5.equivalent\endcsname{0.0}
\expandafter\def\csname val@rev.desc.d1.uiv.p1.C.zoomclick-d3.tau-target.tost5.p\endcsname{0.257}
\expandafter\def\csname cnt@rev.desc.d1.uiv.p1.C.zoomclick-d3.tau-target.tost5.p\endcsname{0}
\expandafter\def\csname pct@rev.desc.d1.uiv.p1.C.zoomclick-d3.tau-target.tost5.p\endcsname{25.7}
\expandafter\def\csname ppt@rev.desc.d1.uiv.p1.C.zoomclick-d3.tau-target.tost5.p\endcsname{$+$25.7}
\expandafter\def\csname val@rev.desc.d1.uiv.p1.C.zoomclick-d3.zero-estimate\endcsname{0}
\expandafter\def\csname cnt@rev.desc.d1.uiv.p1.C.zoomclick-d3.zero-estimate\endcsname{0}
\expandafter\def\csname pct@rev.desc.d1.uiv.p1.C.zoomclick-d3.zero-estimate\endcsname{0.0}
\expandafter\def\csname ppt@rev.desc.d1.uiv.p1.C.zoomclick-d3.zero-estimate\endcsname{0.0}
\expandafter\def\csname val@rev.desc.d1.uiv.p1.Tsealed.center-d1.delta-tau-cross.ci90.0\endcsname{-0.062}
\expandafter\def\csname cnt@rev.desc.d1.uiv.p1.Tsealed.center-d1.delta-tau-cross.ci90.0\endcsname{0}
\expandafter\def\csname pct@rev.desc.d1.uiv.p1.Tsealed.center-d1.delta-tau-cross.ci90.0\endcsname{$-$6.2}
\expandafter\def\csname ppt@rev.desc.d1.uiv.p1.Tsealed.center-d1.delta-tau-cross.ci90.0\endcsname{$-$6.2}
\expandafter\def\csname val@rev.desc.d1.uiv.p1.Tsealed.center-d1.delta-tau-cross.ci90.1\endcsname{0.094}
\expandafter\def\csname cnt@rev.desc.d1.uiv.p1.Tsealed.center-d1.delta-tau-cross.ci90.1\endcsname{0}
\expandafter\def\csname pct@rev.desc.d1.uiv.p1.Tsealed.center-d1.delta-tau-cross.ci90.1\endcsname{9.4}
\expandafter\def\csname ppt@rev.desc.d1.uiv.p1.Tsealed.center-d1.delta-tau-cross.ci90.1\endcsname{$+$9.4}
\expandafter\def\csname val@rev.desc.d1.uiv.p1.Tsealed.center-d1.delta-tau-cross.ci95-excludes-0\endcsname{0}
\expandafter\def\csname cnt@rev.desc.d1.uiv.p1.Tsealed.center-d1.delta-tau-cross.ci95-excludes-0\endcsname{0}
\expandafter\def\csname pct@rev.desc.d1.uiv.p1.Tsealed.center-d1.delta-tau-cross.ci95-excludes-0\endcsname{0.0}
\expandafter\def\csname ppt@rev.desc.d1.uiv.p1.Tsealed.center-d1.delta-tau-cross.ci95-excludes-0\endcsname{0.0}
\expandafter\def\csname val@rev.desc.d1.uiv.p1.Tsealed.center-d1.delta-tau-cross.ci95.0\endcsname{-0.076}
\expandafter\def\csname cnt@rev.desc.d1.uiv.p1.Tsealed.center-d1.delta-tau-cross.ci95.0\endcsname{0}
\expandafter\def\csname pct@rev.desc.d1.uiv.p1.Tsealed.center-d1.delta-tau-cross.ci95.0\endcsname{$-$7.6}
\expandafter\def\csname ppt@rev.desc.d1.uiv.p1.Tsealed.center-d1.delta-tau-cross.ci95.0\endcsname{$-$7.6}
\expandafter\def\csname val@rev.desc.d1.uiv.p1.Tsealed.center-d1.delta-tau-cross.ci95.1\endcsname{0.111}
\expandafter\def\csname cnt@rev.desc.d1.uiv.p1.Tsealed.center-d1.delta-tau-cross.ci95.1\endcsname{0}
\expandafter\def\csname pct@rev.desc.d1.uiv.p1.Tsealed.center-d1.delta-tau-cross.ci95.1\endcsname{11.1}
\expandafter\def\csname ppt@rev.desc.d1.uiv.p1.Tsealed.center-d1.delta-tau-cross.ci95.1\endcsname{$+$11.1}
\expandafter\def\csname val@rev.desc.d1.uiv.p1.Tsealed.center-d1.delta-tau-cross.mean\endcsname{0.017}
\expandafter\def\csname cnt@rev.desc.d1.uiv.p1.Tsealed.center-d1.delta-tau-cross.mean\endcsname{0}
\expandafter\def\csname pct@rev.desc.d1.uiv.p1.Tsealed.center-d1.delta-tau-cross.mean\endcsname{1.7}
\expandafter\def\csname ppt@rev.desc.d1.uiv.p1.Tsealed.center-d1.delta-tau-cross.mean\endcsname{$+$1.7}
\expandafter\def\csname val@rev.desc.d1.uiv.p1.Tsealed.center-d1.delta-tau-cross.n-apps.0\endcsname{24}
\expandafter\def\csname cnt@rev.desc.d1.uiv.p1.Tsealed.center-d1.delta-tau-cross.n-apps.0\endcsname{24}
\expandafter\def\csname pct@rev.desc.d1.uiv.p1.Tsealed.center-d1.delta-tau-cross.n-apps.0\endcsname{2400.0}
\expandafter\def\csname ppt@rev.desc.d1.uiv.p1.Tsealed.center-d1.delta-tau-cross.n-apps.0\endcsname{$+$2400.0}
\expandafter\def\csname val@rev.desc.d1.uiv.p1.Tsealed.center-d1.delta-tau-cross.n-apps.1\endcsname{12}
\expandafter\def\csname cnt@rev.desc.d1.uiv.p1.Tsealed.center-d1.delta-tau-cross.n-apps.1\endcsname{12}
\expandafter\def\csname pct@rev.desc.d1.uiv.p1.Tsealed.center-d1.delta-tau-cross.n-apps.1\endcsname{1200.0}
\expandafter\def\csname ppt@rev.desc.d1.uiv.p1.Tsealed.center-d1.delta-tau-cross.n-apps.1\endcsname{$+$1200.0}
\expandafter\def\csname val@rev.desc.d1.uiv.p1.Tsealed.center-d1.delta-tau-cross.tost1.ci90-inside\endcsname{0}
\expandafter\def\csname cnt@rev.desc.d1.uiv.p1.Tsealed.center-d1.delta-tau-cross.tost1.ci90-inside\endcsname{0}
\expandafter\def\csname pct@rev.desc.d1.uiv.p1.Tsealed.center-d1.delta-tau-cross.tost1.ci90-inside\endcsname{0.0}
\expandafter\def\csname ppt@rev.desc.d1.uiv.p1.Tsealed.center-d1.delta-tau-cross.tost1.ci90-inside\endcsname{0.0}
\expandafter\def\csname val@rev.desc.d1.uiv.p1.Tsealed.center-d1.delta-tau-cross.tost1.equivalent\endcsname{0}
\expandafter\def\csname cnt@rev.desc.d1.uiv.p1.Tsealed.center-d1.delta-tau-cross.tost1.equivalent\endcsname{0}
\expandafter\def\csname pct@rev.desc.d1.uiv.p1.Tsealed.center-d1.delta-tau-cross.tost1.equivalent\endcsname{0.0}
\expandafter\def\csname ppt@rev.desc.d1.uiv.p1.Tsealed.center-d1.delta-tau-cross.tost1.equivalent\endcsname{0.0}
\expandafter\def\csname val@rev.desc.d1.uiv.p1.Tsealed.center-d1.delta-tau-cross.tost1.p\endcsname{0.567}
\expandafter\def\csname cnt@rev.desc.d1.uiv.p1.Tsealed.center-d1.delta-tau-cross.tost1.p\endcsname{1}
\expandafter\def\csname pct@rev.desc.d1.uiv.p1.Tsealed.center-d1.delta-tau-cross.tost1.p\endcsname{56.7}
\expandafter\def\csname ppt@rev.desc.d1.uiv.p1.Tsealed.center-d1.delta-tau-cross.tost1.p\endcsname{$+$56.7}
\expandafter\def\csname val@rev.desc.d1.uiv.p1.Tsealed.center-d1.delta-tau-cross.tost5.ci90-inside\endcsname{0}
\expandafter\def\csname cnt@rev.desc.d1.uiv.p1.Tsealed.center-d1.delta-tau-cross.tost5.ci90-inside\endcsname{0}
\expandafter\def\csname pct@rev.desc.d1.uiv.p1.Tsealed.center-d1.delta-tau-cross.tost5.ci90-inside\endcsname{0.0}
\expandafter\def\csname ppt@rev.desc.d1.uiv.p1.Tsealed.center-d1.delta-tau-cross.tost5.ci90-inside\endcsname{0.0}
\expandafter\def\csname val@rev.desc.d1.uiv.p1.Tsealed.center-d1.delta-tau-cross.tost5.equivalent\endcsname{0}
\expandafter\def\csname cnt@rev.desc.d1.uiv.p1.Tsealed.center-d1.delta-tau-cross.tost5.equivalent\endcsname{0}
\expandafter\def\csname pct@rev.desc.d1.uiv.p1.Tsealed.center-d1.delta-tau-cross.tost5.equivalent\endcsname{0.0}
\expandafter\def\csname ppt@rev.desc.d1.uiv.p1.Tsealed.center-d1.delta-tau-cross.tost5.equivalent\endcsname{0.0}
\expandafter\def\csname val@rev.desc.d1.uiv.p1.Tsealed.center-d1.delta-tau-cross.tost5.p\endcsname{0.240}
\expandafter\def\csname cnt@rev.desc.d1.uiv.p1.Tsealed.center-d1.delta-tau-cross.tost5.p\endcsname{0}
\expandafter\def\csname pct@rev.desc.d1.uiv.p1.Tsealed.center-d1.delta-tau-cross.tost5.p\endcsname{24.0}
\expandafter\def\csname ppt@rev.desc.d1.uiv.p1.Tsealed.center-d1.delta-tau-cross.tost5.p\endcsname{$+$24.0}
\expandafter\def\csname val@rev.desc.d1.uiv.p1.Tsealed.center-d1.material-flip\endcsname{0}
\expandafter\def\csname cnt@rev.desc.d1.uiv.p1.Tsealed.center-d1.material-flip\endcsname{0}
\expandafter\def\csname pct@rev.desc.d1.uiv.p1.Tsealed.center-d1.material-flip\endcsname{0.0}
\expandafter\def\csname ppt@rev.desc.d1.uiv.p1.Tsealed.center-d1.material-flip\endcsname{0.0}
\expandafter\def\csname val@rev.desc.d1.uiv.p1.Tsealed.center-d1.n-tasks\endcsname{144}
\expandafter\def\csname cnt@rev.desc.d1.uiv.p1.Tsealed.center-d1.n-tasks\endcsname{144}
\expandafter\def\csname pct@rev.desc.d1.uiv.p1.Tsealed.center-d1.n-tasks\endcsname{14400.0}
\expandafter\def\csname ppt@rev.desc.d1.uiv.p1.Tsealed.center-d1.n-tasks\endcsname{$+$14400.0}
\expandafter\def\csname val@rev.desc.d1.uiv.p1.Tsealed.center-d1.n-tasks-source-C\endcsname{96}
\expandafter\def\csname cnt@rev.desc.d1.uiv.p1.Tsealed.center-d1.n-tasks-source-C\endcsname{96}
\expandafter\def\csname pct@rev.desc.d1.uiv.p1.Tsealed.center-d1.n-tasks-source-C\endcsname{9600.0}
\expandafter\def\csname ppt@rev.desc.d1.uiv.p1.Tsealed.center-d1.n-tasks-source-C\endcsname{$+$9600.0}
\expandafter\def\csname val@rev.desc.d1.uiv.p1.Tsealed.center-d1.paper-tau-T\endcsname{-0.056}
\expandafter\def\csname cnt@rev.desc.d1.uiv.p1.Tsealed.center-d1.paper-tau-T\endcsname{0}
\expandafter\def\csname pct@rev.desc.d1.uiv.p1.Tsealed.center-d1.paper-tau-T\endcsname{$-$5.6}
\expandafter\def\csname ppt@rev.desc.d1.uiv.p1.Tsealed.center-d1.paper-tau-T\endcsname{$-$5.6}
\expandafter\def\csname val@rev.desc.d1.uiv.p1.Tsealed.center-d1.reproduces\endcsname{1}
\expandafter\def\csname cnt@rev.desc.d1.uiv.p1.Tsealed.center-d1.reproduces\endcsname{1}
\expandafter\def\csname pct@rev.desc.d1.uiv.p1.Tsealed.center-d1.reproduces\endcsname{100.0}
\expandafter\def\csname ppt@rev.desc.d1.uiv.p1.Tsealed.center-d1.reproduces\endcsname{$+$100.0}
\expandafter\def\csname val@rev.desc.d1.uiv.p1.Tsealed.center-d1.reproduces-ci\endcsname{1}
\expandafter\def\csname cnt@rev.desc.d1.uiv.p1.Tsealed.center-d1.reproduces-ci\endcsname{1}
\expandafter\def\csname pct@rev.desc.d1.uiv.p1.Tsealed.center-d1.reproduces-ci\endcsname{100.0}
\expandafter\def\csname ppt@rev.desc.d1.uiv.p1.Tsealed.center-d1.reproduces-ci\endcsname{$+$100.0}
\expandafter\def\csname val@rev.desc.d1.uiv.p1.Tsealed.center-d1.sign-differs\endcsname{0}
\expandafter\def\csname cnt@rev.desc.d1.uiv.p1.Tsealed.center-d1.sign-differs\endcsname{0}
\expandafter\def\csname pct@rev.desc.d1.uiv.p1.Tsealed.center-d1.sign-differs\endcsname{0.0}
\expandafter\def\csname ppt@rev.desc.d1.uiv.p1.Tsealed.center-d1.sign-differs\endcsname{0.0}
\expandafter\def\csname val@rev.desc.d1.uiv.p1.Tsealed.center-d1.tau-source-C.ci90.0\endcsname{-0.135}
\expandafter\def\csname cnt@rev.desc.d1.uiv.p1.Tsealed.center-d1.tau-source-C.ci90.0\endcsname{0}
\expandafter\def\csname pct@rev.desc.d1.uiv.p1.Tsealed.center-d1.tau-source-C.ci90.0\endcsname{$-$13.5}
\expandafter\def\csname ppt@rev.desc.d1.uiv.p1.Tsealed.center-d1.tau-source-C.ci90.0\endcsname{$-$13.5}
\expandafter\def\csname val@rev.desc.d1.uiv.p1.Tsealed.center-d1.tau-source-C.ci90.1\endcsname{-0.010}
\expandafter\def\csname cnt@rev.desc.d1.uiv.p1.Tsealed.center-d1.tau-source-C.ci90.1\endcsname{0}
\expandafter\def\csname pct@rev.desc.d1.uiv.p1.Tsealed.center-d1.tau-source-C.ci90.1\endcsname{$-$1.0}
\expandafter\def\csname ppt@rev.desc.d1.uiv.p1.Tsealed.center-d1.tau-source-C.ci90.1\endcsname{$-$1.0}
\expandafter\def\csname val@rev.desc.d1.uiv.p1.Tsealed.center-d1.tau-source-C.ci95-excludes-0\endcsname{0}
\expandafter\def\csname cnt@rev.desc.d1.uiv.p1.Tsealed.center-d1.tau-source-C.ci95-excludes-0\endcsname{0}
\expandafter\def\csname pct@rev.desc.d1.uiv.p1.Tsealed.center-d1.tau-source-C.ci95-excludes-0\endcsname{0.0}
\expandafter\def\csname ppt@rev.desc.d1.uiv.p1.Tsealed.center-d1.tau-source-C.ci95-excludes-0\endcsname{0.0}
\expandafter\def\csname val@rev.desc.d1.uiv.p1.Tsealed.center-d1.tau-source-C.ci95.0\endcsname{-0.146}
\expandafter\def\csname cnt@rev.desc.d1.uiv.p1.Tsealed.center-d1.tau-source-C.ci95.0\endcsname{0}
\expandafter\def\csname pct@rev.desc.d1.uiv.p1.Tsealed.center-d1.tau-source-C.ci95.0\endcsname{$-$14.6}
\expandafter\def\csname ppt@rev.desc.d1.uiv.p1.Tsealed.center-d1.tau-source-C.ci95.0\endcsname{$-$14.6}
\expandafter\def\csname val@rev.desc.d1.uiv.p1.Tsealed.center-d1.tau-source-C.ci95.1\endcsname{0.000}
\expandafter\def\csname cnt@rev.desc.d1.uiv.p1.Tsealed.center-d1.tau-source-C.ci95.1\endcsname{0}
\expandafter\def\csname pct@rev.desc.d1.uiv.p1.Tsealed.center-d1.tau-source-C.ci95.1\endcsname{0.0}
\expandafter\def\csname ppt@rev.desc.d1.uiv.p1.Tsealed.center-d1.tau-source-C.ci95.1\endcsname{0.0}
\expandafter\def\csname val@rev.desc.d1.uiv.p1.Tsealed.center-d1.tau-source-C.mean\endcsname{-0.073}
\expandafter\def\csname cnt@rev.desc.d1.uiv.p1.Tsealed.center-d1.tau-source-C.mean\endcsname{0}
\expandafter\def\csname pct@rev.desc.d1.uiv.p1.Tsealed.center-d1.tau-source-C.mean\endcsname{$-$7.3}
\expandafter\def\csname ppt@rev.desc.d1.uiv.p1.Tsealed.center-d1.tau-source-C.mean\endcsname{$-$7.3}
\expandafter\def\csname val@rev.desc.d1.uiv.p1.Tsealed.center-d1.tau-source-C.n-apps\endcsname{12}
\expandafter\def\csname cnt@rev.desc.d1.uiv.p1.Tsealed.center-d1.tau-source-C.n-apps\endcsname{12}
\expandafter\def\csname pct@rev.desc.d1.uiv.p1.Tsealed.center-d1.tau-source-C.n-apps\endcsname{1200.0}
\expandafter\def\csname ppt@rev.desc.d1.uiv.p1.Tsealed.center-d1.tau-source-C.n-apps\endcsname{$+$1200.0}
\expandafter\def\csname val@rev.desc.d1.uiv.p1.Tsealed.center-d1.tau-source-C.tost1.ci90-inside\endcsname{0}
\expandafter\def\csname cnt@rev.desc.d1.uiv.p1.Tsealed.center-d1.tau-source-C.tost1.ci90-inside\endcsname{0}
\expandafter\def\csname pct@rev.desc.d1.uiv.p1.Tsealed.center-d1.tau-source-C.tost1.ci90-inside\endcsname{0.0}
\expandafter\def\csname ppt@rev.desc.d1.uiv.p1.Tsealed.center-d1.tau-source-C.tost1.ci90-inside\endcsname{0.0}
\expandafter\def\csname val@rev.desc.d1.uiv.p1.Tsealed.center-d1.tau-source-C.tost1.equivalent\endcsname{0}
\expandafter\def\csname cnt@rev.desc.d1.uiv.p1.Tsealed.center-d1.tau-source-C.tost1.equivalent\endcsname{0}
\expandafter\def\csname pct@rev.desc.d1.uiv.p1.Tsealed.center-d1.tau-source-C.tost1.equivalent\endcsname{0.0}
\expandafter\def\csname ppt@rev.desc.d1.uiv.p1.Tsealed.center-d1.tau-source-C.tost1.equivalent\endcsname{0.0}
\expandafter\def\csname val@rev.desc.d1.uiv.p1.Tsealed.center-d1.tau-source-C.tost1.p\endcsname{0.970}
\expandafter\def\csname cnt@rev.desc.d1.uiv.p1.Tsealed.center-d1.tau-source-C.tost1.p\endcsname{1}
\expandafter\def\csname pct@rev.desc.d1.uiv.p1.Tsealed.center-d1.tau-source-C.tost1.p\endcsname{97.0}
\expandafter\def\csname ppt@rev.desc.d1.uiv.p1.Tsealed.center-d1.tau-source-C.tost1.p\endcsname{$+$97.0}
\expandafter\def\csname val@rev.desc.d1.uiv.p1.Tsealed.center-d1.tau-source-C.tost5.ci90-inside\endcsname{0}
\expandafter\def\csname cnt@rev.desc.d1.uiv.p1.Tsealed.center-d1.tau-source-C.tost5.ci90-inside\endcsname{0}
\expandafter\def\csname pct@rev.desc.d1.uiv.p1.Tsealed.center-d1.tau-source-C.tost5.ci90-inside\endcsname{0.0}
\expandafter\def\csname ppt@rev.desc.d1.uiv.p1.Tsealed.center-d1.tau-source-C.tost5.ci90-inside\endcsname{0.0}
\expandafter\def\csname val@rev.desc.d1.uiv.p1.Tsealed.center-d1.tau-source-C.tost5.equivalent\endcsname{0}
\expandafter\def\csname cnt@rev.desc.d1.uiv.p1.Tsealed.center-d1.tau-source-C.tost5.equivalent\endcsname{0}
\expandafter\def\csname pct@rev.desc.d1.uiv.p1.Tsealed.center-d1.tau-source-C.tost5.equivalent\endcsname{0.0}
\expandafter\def\csname ppt@rev.desc.d1.uiv.p1.Tsealed.center-d1.tau-source-C.tost5.equivalent\endcsname{0.0}
\expandafter\def\csname val@rev.desc.d1.uiv.p1.Tsealed.center-d1.tau-source-C.tost5.p\endcsname{0.753}
\expandafter\def\csname cnt@rev.desc.d1.uiv.p1.Tsealed.center-d1.tau-source-C.tost5.p\endcsname{1}
\expandafter\def\csname pct@rev.desc.d1.uiv.p1.Tsealed.center-d1.tau-source-C.tost5.p\endcsname{75.3}
\expandafter\def\csname ppt@rev.desc.d1.uiv.p1.Tsealed.center-d1.tau-source-C.tost5.p\endcsname{$+$75.3}
\expandafter\def\csname val@rev.desc.d1.uiv.p1.Tsealed.center-d1.tau-target-T.ci90.0\endcsname{-0.111}
\expandafter\def\csname cnt@rev.desc.d1.uiv.p1.Tsealed.center-d1.tau-target-T.ci90.0\endcsname{0}
\expandafter\def\csname pct@rev.desc.d1.uiv.p1.Tsealed.center-d1.tau-target-T.ci90.0\endcsname{$-$11.1}
\expandafter\def\csname ppt@rev.desc.d1.uiv.p1.Tsealed.center-d1.tau-target-T.ci90.0\endcsname{$-$11.1}
\expandafter\def\csname val@rev.desc.d1.uiv.p1.Tsealed.center-d1.tau-target-T.ci90.1\endcsname{-0.007}
\expandafter\def\csname cnt@rev.desc.d1.uiv.p1.Tsealed.center-d1.tau-target-T.ci90.1\endcsname{0}
\expandafter\def\csname pct@rev.desc.d1.uiv.p1.Tsealed.center-d1.tau-target-T.ci90.1\endcsname{$-$0.7}
\expandafter\def\csname ppt@rev.desc.d1.uiv.p1.Tsealed.center-d1.tau-target-T.ci90.1\endcsname{$-$0.7}
\expandafter\def\csname val@rev.desc.d1.uiv.p1.Tsealed.center-d1.tau-target-T.ci95-excludes-0\endcsname{1}
\expandafter\def\csname cnt@rev.desc.d1.uiv.p1.Tsealed.center-d1.tau-target-T.ci95-excludes-0\endcsname{1}
\expandafter\def\csname pct@rev.desc.d1.uiv.p1.Tsealed.center-d1.tau-target-T.ci95-excludes-0\endcsname{100.0}
\expandafter\def\csname ppt@rev.desc.d1.uiv.p1.Tsealed.center-d1.tau-target-T.ci95-excludes-0\endcsname{$+$100.0}
\expandafter\def\csname val@rev.desc.d1.uiv.p1.Tsealed.center-d1.tau-target-T.ci95.0\endcsname{-0.118}
\expandafter\def\csname cnt@rev.desc.d1.uiv.p1.Tsealed.center-d1.tau-target-T.ci95.0\endcsname{0}
\expandafter\def\csname pct@rev.desc.d1.uiv.p1.Tsealed.center-d1.tau-target-T.ci95.0\endcsname{$-$11.8}
\expandafter\def\csname ppt@rev.desc.d1.uiv.p1.Tsealed.center-d1.tau-target-T.ci95.0\endcsname{$-$11.8}
\expandafter\def\csname val@rev.desc.d1.uiv.p1.Tsealed.center-d1.tau-target-T.ci95.1\endcsname{-0.000}
\expandafter\def\csname cnt@rev.desc.d1.uiv.p1.Tsealed.center-d1.tau-target-T.ci95.1\endcsname{0}
\expandafter\def\csname pct@rev.desc.d1.uiv.p1.Tsealed.center-d1.tau-target-T.ci95.1\endcsname{0.0}
\expandafter\def\csname ppt@rev.desc.d1.uiv.p1.Tsealed.center-d1.tau-target-T.ci95.1\endcsname{0.0}
\expandafter\def\csname val@rev.desc.d1.uiv.p1.Tsealed.center-d1.tau-target-T.mean\endcsname{-0.056}
\expandafter\def\csname cnt@rev.desc.d1.uiv.p1.Tsealed.center-d1.tau-target-T.mean\endcsname{0}
\expandafter\def\csname pct@rev.desc.d1.uiv.p1.Tsealed.center-d1.tau-target-T.mean\endcsname{$-$5.6}
\expandafter\def\csname ppt@rev.desc.d1.uiv.p1.Tsealed.center-d1.tau-target-T.mean\endcsname{$-$5.6}
\expandafter\def\csname val@rev.desc.d1.uiv.p1.Tsealed.center-d1.tau-target-T.n-apps\endcsname{24}
\expandafter\def\csname cnt@rev.desc.d1.uiv.p1.Tsealed.center-d1.tau-target-T.n-apps\endcsname{24}
\expandafter\def\csname pct@rev.desc.d1.uiv.p1.Tsealed.center-d1.tau-target-T.n-apps\endcsname{2400.0}
\expandafter\def\csname ppt@rev.desc.d1.uiv.p1.Tsealed.center-d1.tau-target-T.n-apps\endcsname{$+$2400.0}
\expandafter\def\csname val@rev.desc.d1.uiv.p1.Tsealed.center-d1.tau-target-T.tost1.ci90-inside\endcsname{0}
\expandafter\def\csname cnt@rev.desc.d1.uiv.p1.Tsealed.center-d1.tau-target-T.tost1.ci90-inside\endcsname{0}
\expandafter\def\csname pct@rev.desc.d1.uiv.p1.Tsealed.center-d1.tau-target-T.tost1.ci90-inside\endcsname{0.0}
\expandafter\def\csname ppt@rev.desc.d1.uiv.p1.Tsealed.center-d1.tau-target-T.tost1.ci90-inside\endcsname{0.0}
\expandafter\def\csname val@rev.desc.d1.uiv.p1.Tsealed.center-d1.tau-target-T.tost1.equivalent\endcsname{0}
\expandafter\def\csname cnt@rev.desc.d1.uiv.p1.Tsealed.center-d1.tau-target-T.tost1.equivalent\endcsname{0}
\expandafter\def\csname pct@rev.desc.d1.uiv.p1.Tsealed.center-d1.tau-target-T.tost1.equivalent\endcsname{0.0}
\expandafter\def\csname ppt@rev.desc.d1.uiv.p1.Tsealed.center-d1.tau-target-T.tost1.equivalent\endcsname{0.0}
\expandafter\def\csname val@rev.desc.d1.uiv.p1.Tsealed.center-d1.tau-target-T.tost1.p\endcsname{0.939}
\expandafter\def\csname cnt@rev.desc.d1.uiv.p1.Tsealed.center-d1.tau-target-T.tost1.p\endcsname{1}
\expandafter\def\csname pct@rev.desc.d1.uiv.p1.Tsealed.center-d1.tau-target-T.tost1.p\endcsname{93.9}
\expandafter\def\csname ppt@rev.desc.d1.uiv.p1.Tsealed.center-d1.tau-target-T.tost1.p\endcsname{$+$93.9}
\expandafter\def\csname val@rev.desc.d1.uiv.p1.Tsealed.center-d1.tau-target-T.tost5.ci90-inside\endcsname{0}
\expandafter\def\csname cnt@rev.desc.d1.uiv.p1.Tsealed.center-d1.tau-target-T.tost5.ci90-inside\endcsname{0}
\expandafter\def\csname pct@rev.desc.d1.uiv.p1.Tsealed.center-d1.tau-target-T.tost5.ci90-inside\endcsname{0.0}
\expandafter\def\csname ppt@rev.desc.d1.uiv.p1.Tsealed.center-d1.tau-target-T.tost5.ci90-inside\endcsname{0.0}
\expandafter\def\csname val@rev.desc.d1.uiv.p1.Tsealed.center-d1.tau-target-T.tost5.equivalent\endcsname{0}
\expandafter\def\csname cnt@rev.desc.d1.uiv.p1.Tsealed.center-d1.tau-target-T.tost5.equivalent\endcsname{0}
\expandafter\def\csname pct@rev.desc.d1.uiv.p1.Tsealed.center-d1.tau-target-T.tost5.equivalent\endcsname{0.0}
\expandafter\def\csname ppt@rev.desc.d1.uiv.p1.Tsealed.center-d1.tau-target-T.tost5.equivalent\endcsname{0.0}
\expandafter\def\csname val@rev.desc.d1.uiv.p1.Tsealed.center-d1.tau-target-T.tost5.p\endcsname{0.528}
\expandafter\def\csname cnt@rev.desc.d1.uiv.p1.Tsealed.center-d1.tau-target-T.tost5.p\endcsname{1}
\expandafter\def\csname pct@rev.desc.d1.uiv.p1.Tsealed.center-d1.tau-target-T.tost5.p\endcsname{52.8}
\expandafter\def\csname ppt@rev.desc.d1.uiv.p1.Tsealed.center-d1.tau-target-T.tost5.p\endcsname{$+$52.8}
\expandafter\def\csname val@rev.desc.d1.uiv.p1.Tsealed.center-d1.zero-estimate\endcsname{0}
\expandafter\def\csname cnt@rev.desc.d1.uiv.p1.Tsealed.center-d1.zero-estimate\endcsname{0}
\expandafter\def\csname pct@rev.desc.d1.uiv.p1.Tsealed.center-d1.zero-estimate\endcsname{0.0}
\expandafter\def\csname ppt@rev.desc.d1.uiv.p1.Tsealed.center-d1.zero-estimate\endcsname{0.0}
\expandafter\def\csname val@rev.desc.d1.uiv.p1.Tsealed.context-d1.delta-tau-cross.ci90.0\endcsname{-0.052}
\expandafter\def\csname cnt@rev.desc.d1.uiv.p1.Tsealed.context-d1.delta-tau-cross.ci90.0\endcsname{0}
\expandafter\def\csname pct@rev.desc.d1.uiv.p1.Tsealed.context-d1.delta-tau-cross.ci90.0\endcsname{$-$5.2}
\expandafter\def\csname ppt@rev.desc.d1.uiv.p1.Tsealed.context-d1.delta-tau-cross.ci90.0\endcsname{$-$5.2}
\expandafter\def\csname val@rev.desc.d1.uiv.p1.Tsealed.context-d1.delta-tau-cross.ci90.1\endcsname{0.087}
\expandafter\def\csname cnt@rev.desc.d1.uiv.p1.Tsealed.context-d1.delta-tau-cross.ci90.1\endcsname{0}
\expandafter\def\csname pct@rev.desc.d1.uiv.p1.Tsealed.context-d1.delta-tau-cross.ci90.1\endcsname{8.7}
\expandafter\def\csname ppt@rev.desc.d1.uiv.p1.Tsealed.context-d1.delta-tau-cross.ci90.1\endcsname{$+$8.7}
\expandafter\def\csname val@rev.desc.d1.uiv.p1.Tsealed.context-d1.delta-tau-cross.ci95-excludes-0\endcsname{0}
\expandafter\def\csname cnt@rev.desc.d1.uiv.p1.Tsealed.context-d1.delta-tau-cross.ci95-excludes-0\endcsname{0}
\expandafter\def\csname pct@rev.desc.d1.uiv.p1.Tsealed.context-d1.delta-tau-cross.ci95-excludes-0\endcsname{0.0}
\expandafter\def\csname ppt@rev.desc.d1.uiv.p1.Tsealed.context-d1.delta-tau-cross.ci95-excludes-0\endcsname{0.0}
\expandafter\def\csname val@rev.desc.d1.uiv.p1.Tsealed.context-d1.delta-tau-cross.ci95.0\endcsname{-0.066}
\expandafter\def\csname cnt@rev.desc.d1.uiv.p1.Tsealed.context-d1.delta-tau-cross.ci95.0\endcsname{0}
\expandafter\def\csname pct@rev.desc.d1.uiv.p1.Tsealed.context-d1.delta-tau-cross.ci95.0\endcsname{$-$6.6}
\expandafter\def\csname ppt@rev.desc.d1.uiv.p1.Tsealed.context-d1.delta-tau-cross.ci95.0\endcsname{$-$6.6}
\expandafter\def\csname val@rev.desc.d1.uiv.p1.Tsealed.context-d1.delta-tau-cross.ci95.1\endcsname{0.104}
\expandafter\def\csname cnt@rev.desc.d1.uiv.p1.Tsealed.context-d1.delta-tau-cross.ci95.1\endcsname{0}
\expandafter\def\csname pct@rev.desc.d1.uiv.p1.Tsealed.context-d1.delta-tau-cross.ci95.1\endcsname{10.4}
\expandafter\def\csname ppt@rev.desc.d1.uiv.p1.Tsealed.context-d1.delta-tau-cross.ci95.1\endcsname{$+$10.4}
\expandafter\def\csname val@rev.desc.d1.uiv.p1.Tsealed.context-d1.delta-tau-cross.mean\endcsname{0.017}
\expandafter\def\csname cnt@rev.desc.d1.uiv.p1.Tsealed.context-d1.delta-tau-cross.mean\endcsname{0}
\expandafter\def\csname pct@rev.desc.d1.uiv.p1.Tsealed.context-d1.delta-tau-cross.mean\endcsname{1.7}
\expandafter\def\csname ppt@rev.desc.d1.uiv.p1.Tsealed.context-d1.delta-tau-cross.mean\endcsname{$+$1.7}
\expandafter\def\csname val@rev.desc.d1.uiv.p1.Tsealed.context-d1.delta-tau-cross.n-apps.0\endcsname{24}
\expandafter\def\csname cnt@rev.desc.d1.uiv.p1.Tsealed.context-d1.delta-tau-cross.n-apps.0\endcsname{24}
\expandafter\def\csname pct@rev.desc.d1.uiv.p1.Tsealed.context-d1.delta-tau-cross.n-apps.0\endcsname{2400.0}
\expandafter\def\csname ppt@rev.desc.d1.uiv.p1.Tsealed.context-d1.delta-tau-cross.n-apps.0\endcsname{$+$2400.0}
\expandafter\def\csname val@rev.desc.d1.uiv.p1.Tsealed.context-d1.delta-tau-cross.n-apps.1\endcsname{12}
\expandafter\def\csname cnt@rev.desc.d1.uiv.p1.Tsealed.context-d1.delta-tau-cross.n-apps.1\endcsname{12}
\expandafter\def\csname pct@rev.desc.d1.uiv.p1.Tsealed.context-d1.delta-tau-cross.n-apps.1\endcsname{1200.0}
\expandafter\def\csname ppt@rev.desc.d1.uiv.p1.Tsealed.context-d1.delta-tau-cross.n-apps.1\endcsname{$+$1200.0}
\expandafter\def\csname val@rev.desc.d1.uiv.p1.Tsealed.context-d1.delta-tau-cross.tost1.ci90-inside\endcsname{0}
\expandafter\def\csname cnt@rev.desc.d1.uiv.p1.Tsealed.context-d1.delta-tau-cross.tost1.ci90-inside\endcsname{0}
\expandafter\def\csname pct@rev.desc.d1.uiv.p1.Tsealed.context-d1.delta-tau-cross.tost1.ci90-inside\endcsname{0.0}
\expandafter\def\csname ppt@rev.desc.d1.uiv.p1.Tsealed.context-d1.delta-tau-cross.tost1.ci90-inside\endcsname{0.0}
\expandafter\def\csname val@rev.desc.d1.uiv.p1.Tsealed.context-d1.delta-tau-cross.tost1.equivalent\endcsname{0}
\expandafter\def\csname cnt@rev.desc.d1.uiv.p1.Tsealed.context-d1.delta-tau-cross.tost1.equivalent\endcsname{0}
\expandafter\def\csname pct@rev.desc.d1.uiv.p1.Tsealed.context-d1.delta-tau-cross.tost1.equivalent\endcsname{0.0}
\expandafter\def\csname ppt@rev.desc.d1.uiv.p1.Tsealed.context-d1.delta-tau-cross.tost1.equivalent\endcsname{0.0}
\expandafter\def\csname val@rev.desc.d1.uiv.p1.Tsealed.context-d1.delta-tau-cross.tost1.p\endcsname{0.585}
\expandafter\def\csname cnt@rev.desc.d1.uiv.p1.Tsealed.context-d1.delta-tau-cross.tost1.p\endcsname{1}
\expandafter\def\csname pct@rev.desc.d1.uiv.p1.Tsealed.context-d1.delta-tau-cross.tost1.p\endcsname{58.5}
\expandafter\def\csname ppt@rev.desc.d1.uiv.p1.Tsealed.context-d1.delta-tau-cross.tost1.p\endcsname{$+$58.5}
\expandafter\def\csname val@rev.desc.d1.uiv.p1.Tsealed.context-d1.delta-tau-cross.tost5.ci90-inside\endcsname{0}
\expandafter\def\csname cnt@rev.desc.d1.uiv.p1.Tsealed.context-d1.delta-tau-cross.tost5.ci90-inside\endcsname{0}
\expandafter\def\csname pct@rev.desc.d1.uiv.p1.Tsealed.context-d1.delta-tau-cross.tost5.ci90-inside\endcsname{0.0}
\expandafter\def\csname ppt@rev.desc.d1.uiv.p1.Tsealed.context-d1.delta-tau-cross.tost5.ci90-inside\endcsname{0.0}
\expandafter\def\csname val@rev.desc.d1.uiv.p1.Tsealed.context-d1.delta-tau-cross.tost5.equivalent\endcsname{0}
\expandafter\def\csname cnt@rev.desc.d1.uiv.p1.Tsealed.context-d1.delta-tau-cross.tost5.equivalent\endcsname{0}
\expandafter\def\csname pct@rev.desc.d1.uiv.p1.Tsealed.context-d1.delta-tau-cross.tost5.equivalent\endcsname{0.0}
\expandafter\def\csname ppt@rev.desc.d1.uiv.p1.Tsealed.context-d1.delta-tau-cross.tost5.equivalent\endcsname{0.0}
\expandafter\def\csname val@rev.desc.d1.uiv.p1.Tsealed.context-d1.delta-tau-cross.tost5.p\endcsname{0.221}
\expandafter\def\csname cnt@rev.desc.d1.uiv.p1.Tsealed.context-d1.delta-tau-cross.tost5.p\endcsname{0}
\expandafter\def\csname pct@rev.desc.d1.uiv.p1.Tsealed.context-d1.delta-tau-cross.tost5.p\endcsname{22.1}
\expandafter\def\csname ppt@rev.desc.d1.uiv.p1.Tsealed.context-d1.delta-tau-cross.tost5.p\endcsname{$+$22.1}
\expandafter\def\csname val@rev.desc.d1.uiv.p1.Tsealed.context-d1.material-flip\endcsname{0}
\expandafter\def\csname cnt@rev.desc.d1.uiv.p1.Tsealed.context-d1.material-flip\endcsname{0}
\expandafter\def\csname pct@rev.desc.d1.uiv.p1.Tsealed.context-d1.material-flip\endcsname{0.0}
\expandafter\def\csname ppt@rev.desc.d1.uiv.p1.Tsealed.context-d1.material-flip\endcsname{0.0}
\expandafter\def\csname val@rev.desc.d1.uiv.p1.Tsealed.context-d1.n-tasks\endcsname{144}
\expandafter\def\csname cnt@rev.desc.d1.uiv.p1.Tsealed.context-d1.n-tasks\endcsname{144}
\expandafter\def\csname pct@rev.desc.d1.uiv.p1.Tsealed.context-d1.n-tasks\endcsname{14400.0}
\expandafter\def\csname ppt@rev.desc.d1.uiv.p1.Tsealed.context-d1.n-tasks\endcsname{$+$14400.0}
\expandafter\def\csname val@rev.desc.d1.uiv.p1.Tsealed.context-d1.n-tasks-source-C\endcsname{96}
\expandafter\def\csname cnt@rev.desc.d1.uiv.p1.Tsealed.context-d1.n-tasks-source-C\endcsname{96}
\expandafter\def\csname pct@rev.desc.d1.uiv.p1.Tsealed.context-d1.n-tasks-source-C\endcsname{9600.0}
\expandafter\def\csname ppt@rev.desc.d1.uiv.p1.Tsealed.context-d1.n-tasks-source-C\endcsname{$+$9600.0}
\expandafter\def\csname val@rev.desc.d1.uiv.p1.Tsealed.context-d1.paper-tau-T\endcsname{-0.056}
\expandafter\def\csname cnt@rev.desc.d1.uiv.p1.Tsealed.context-d1.paper-tau-T\endcsname{0}
\expandafter\def\csname pct@rev.desc.d1.uiv.p1.Tsealed.context-d1.paper-tau-T\endcsname{$-$5.6}
\expandafter\def\csname ppt@rev.desc.d1.uiv.p1.Tsealed.context-d1.paper-tau-T\endcsname{$-$5.6}
\expandafter\def\csname val@rev.desc.d1.uiv.p1.Tsealed.context-d1.reproduces\endcsname{1}
\expandafter\def\csname cnt@rev.desc.d1.uiv.p1.Tsealed.context-d1.reproduces\endcsname{1}
\expandafter\def\csname pct@rev.desc.d1.uiv.p1.Tsealed.context-d1.reproduces\endcsname{100.0}
\expandafter\def\csname ppt@rev.desc.d1.uiv.p1.Tsealed.context-d1.reproduces\endcsname{$+$100.0}
\expandafter\def\csname val@rev.desc.d1.uiv.p1.Tsealed.context-d1.reproduces-ci\endcsname{1}
\expandafter\def\csname cnt@rev.desc.d1.uiv.p1.Tsealed.context-d1.reproduces-ci\endcsname{1}
\expandafter\def\csname pct@rev.desc.d1.uiv.p1.Tsealed.context-d1.reproduces-ci\endcsname{100.0}
\expandafter\def\csname ppt@rev.desc.d1.uiv.p1.Tsealed.context-d1.reproduces-ci\endcsname{$+$100.0}
\expandafter\def\csname val@rev.desc.d1.uiv.p1.Tsealed.context-d1.sign-differs\endcsname{0}
\expandafter\def\csname cnt@rev.desc.d1.uiv.p1.Tsealed.context-d1.sign-differs\endcsname{0}
\expandafter\def\csname pct@rev.desc.d1.uiv.p1.Tsealed.context-d1.sign-differs\endcsname{0.0}
\expandafter\def\csname ppt@rev.desc.d1.uiv.p1.Tsealed.context-d1.sign-differs\endcsname{0.0}
\expandafter\def\csname val@rev.desc.d1.uiv.p1.Tsealed.context-d1.tau-source-C.ci90.0\endcsname{-0.135}
\expandafter\def\csname cnt@rev.desc.d1.uiv.p1.Tsealed.context-d1.tau-source-C.ci90.0\endcsname{0}
\expandafter\def\csname pct@rev.desc.d1.uiv.p1.Tsealed.context-d1.tau-source-C.ci90.0\endcsname{$-$13.5}
\expandafter\def\csname ppt@rev.desc.d1.uiv.p1.Tsealed.context-d1.tau-source-C.ci90.0\endcsname{$-$13.5}
\expandafter\def\csname val@rev.desc.d1.uiv.p1.Tsealed.context-d1.tau-source-C.ci90.1\endcsname{-0.021}
\expandafter\def\csname cnt@rev.desc.d1.uiv.p1.Tsealed.context-d1.tau-source-C.ci90.1\endcsname{0}
\expandafter\def\csname pct@rev.desc.d1.uiv.p1.Tsealed.context-d1.tau-source-C.ci90.1\endcsname{$-$2.1}
\expandafter\def\csname ppt@rev.desc.d1.uiv.p1.Tsealed.context-d1.tau-source-C.ci90.1\endcsname{$-$2.1}
\expandafter\def\csname val@rev.desc.d1.uiv.p1.Tsealed.context-d1.tau-source-C.ci95-excludes-0\endcsname{1}
\expandafter\def\csname cnt@rev.desc.d1.uiv.p1.Tsealed.context-d1.tau-source-C.ci95-excludes-0\endcsname{1}
\expandafter\def\csname pct@rev.desc.d1.uiv.p1.Tsealed.context-d1.tau-source-C.ci95-excludes-0\endcsname{100.0}
\expandafter\def\csname ppt@rev.desc.d1.uiv.p1.Tsealed.context-d1.tau-source-C.ci95-excludes-0\endcsname{$+$100.0}
\expandafter\def\csname val@rev.desc.d1.uiv.p1.Tsealed.context-d1.tau-source-C.ci95.0\endcsname{-0.146}
\expandafter\def\csname cnt@rev.desc.d1.uiv.p1.Tsealed.context-d1.tau-source-C.ci95.0\endcsname{0}
\expandafter\def\csname pct@rev.desc.d1.uiv.p1.Tsealed.context-d1.tau-source-C.ci95.0\endcsname{$-$14.6}
\expandafter\def\csname ppt@rev.desc.d1.uiv.p1.Tsealed.context-d1.tau-source-C.ci95.0\endcsname{$-$14.6}
\expandafter\def\csname val@rev.desc.d1.uiv.p1.Tsealed.context-d1.tau-source-C.ci95.1\endcsname{-0.010}
\expandafter\def\csname cnt@rev.desc.d1.uiv.p1.Tsealed.context-d1.tau-source-C.ci95.1\endcsname{0}
\expandafter\def\csname pct@rev.desc.d1.uiv.p1.Tsealed.context-d1.tau-source-C.ci95.1\endcsname{$-$1.0}
\expandafter\def\csname ppt@rev.desc.d1.uiv.p1.Tsealed.context-d1.tau-source-C.ci95.1\endcsname{$-$1.0}
\expandafter\def\csname val@rev.desc.d1.uiv.p1.Tsealed.context-d1.tau-source-C.mean\endcsname{-0.073}
\expandafter\def\csname cnt@rev.desc.d1.uiv.p1.Tsealed.context-d1.tau-source-C.mean\endcsname{0}
\expandafter\def\csname pct@rev.desc.d1.uiv.p1.Tsealed.context-d1.tau-source-C.mean\endcsname{$-$7.3}
\expandafter\def\csname ppt@rev.desc.d1.uiv.p1.Tsealed.context-d1.tau-source-C.mean\endcsname{$-$7.3}
\expandafter\def\csname val@rev.desc.d1.uiv.p1.Tsealed.context-d1.tau-source-C.n-apps\endcsname{12}
\expandafter\def\csname cnt@rev.desc.d1.uiv.p1.Tsealed.context-d1.tau-source-C.n-apps\endcsname{12}
\expandafter\def\csname pct@rev.desc.d1.uiv.p1.Tsealed.context-d1.tau-source-C.n-apps\endcsname{1200.0}
\expandafter\def\csname ppt@rev.desc.d1.uiv.p1.Tsealed.context-d1.tau-source-C.n-apps\endcsname{$+$1200.0}
\expandafter\def\csname val@rev.desc.d1.uiv.p1.Tsealed.context-d1.tau-source-C.tost1.ci90-inside\endcsname{0}
\expandafter\def\csname cnt@rev.desc.d1.uiv.p1.Tsealed.context-d1.tau-source-C.tost1.ci90-inside\endcsname{0}
\expandafter\def\csname pct@rev.desc.d1.uiv.p1.Tsealed.context-d1.tau-source-C.tost1.ci90-inside\endcsname{0.0}
\expandafter\def\csname ppt@rev.desc.d1.uiv.p1.Tsealed.context-d1.tau-source-C.tost1.ci90-inside\endcsname{0.0}
\expandafter\def\csname val@rev.desc.d1.uiv.p1.Tsealed.context-d1.tau-source-C.tost1.equivalent\endcsname{0}
\expandafter\def\csname cnt@rev.desc.d1.uiv.p1.Tsealed.context-d1.tau-source-C.tost1.equivalent\endcsname{0}
\expandafter\def\csname pct@rev.desc.d1.uiv.p1.Tsealed.context-d1.tau-source-C.tost1.equivalent\endcsname{0.0}
\expandafter\def\csname ppt@rev.desc.d1.uiv.p1.Tsealed.context-d1.tau-source-C.tost1.equivalent\endcsname{0.0}
\expandafter\def\csname val@rev.desc.d1.uiv.p1.Tsealed.context-d1.tau-source-C.tost1.p\endcsname{0.976}
\expandafter\def\csname cnt@rev.desc.d1.uiv.p1.Tsealed.context-d1.tau-source-C.tost1.p\endcsname{1}
\expandafter\def\csname pct@rev.desc.d1.uiv.p1.Tsealed.context-d1.tau-source-C.tost1.p\endcsname{97.6}
\expandafter\def\csname ppt@rev.desc.d1.uiv.p1.Tsealed.context-d1.tau-source-C.tost1.p\endcsname{$+$97.6}
\expandafter\def\csname val@rev.desc.d1.uiv.p1.Tsealed.context-d1.tau-source-C.tost5.ci90-inside\endcsname{0}
\expandafter\def\csname cnt@rev.desc.d1.uiv.p1.Tsealed.context-d1.tau-source-C.tost5.ci90-inside\endcsname{0}
\expandafter\def\csname pct@rev.desc.d1.uiv.p1.Tsealed.context-d1.tau-source-C.tost5.ci90-inside\endcsname{0.0}
\expandafter\def\csname ppt@rev.desc.d1.uiv.p1.Tsealed.context-d1.tau-source-C.tost5.ci90-inside\endcsname{0.0}
\expandafter\def\csname val@rev.desc.d1.uiv.p1.Tsealed.context-d1.tau-source-C.tost5.equivalent\endcsname{0}
\expandafter\def\csname cnt@rev.desc.d1.uiv.p1.Tsealed.context-d1.tau-source-C.tost5.equivalent\endcsname{0}
\expandafter\def\csname pct@rev.desc.d1.uiv.p1.Tsealed.context-d1.tau-source-C.tost5.equivalent\endcsname{0.0}
\expandafter\def\csname ppt@rev.desc.d1.uiv.p1.Tsealed.context-d1.tau-source-C.tost5.equivalent\endcsname{0.0}
\expandafter\def\csname val@rev.desc.d1.uiv.p1.Tsealed.context-d1.tau-source-C.tost5.p\endcsname{0.763}
\expandafter\def\csname cnt@rev.desc.d1.uiv.p1.Tsealed.context-d1.tau-source-C.tost5.p\endcsname{1}
\expandafter\def\csname pct@rev.desc.d1.uiv.p1.Tsealed.context-d1.tau-source-C.tost5.p\endcsname{76.3}
\expandafter\def\csname ppt@rev.desc.d1.uiv.p1.Tsealed.context-d1.tau-source-C.tost5.p\endcsname{$+$76.3}
\expandafter\def\csname val@rev.desc.d1.uiv.p1.Tsealed.context-d1.tau-target-T.ci90.0\endcsname{-0.104}
\expandafter\def\csname cnt@rev.desc.d1.uiv.p1.Tsealed.context-d1.tau-target-T.ci90.0\endcsname{0}
\expandafter\def\csname pct@rev.desc.d1.uiv.p1.Tsealed.context-d1.tau-target-T.ci90.0\endcsname{$-$10.4}
\expandafter\def\csname ppt@rev.desc.d1.uiv.p1.Tsealed.context-d1.tau-target-T.ci90.0\endcsname{$-$10.4}
\expandafter\def\csname val@rev.desc.d1.uiv.p1.Tsealed.context-d1.tau-target-T.ci90.1\endcsname{-0.014}
\expandafter\def\csname cnt@rev.desc.d1.uiv.p1.Tsealed.context-d1.tau-target-T.ci90.1\endcsname{0}
\expandafter\def\csname pct@rev.desc.d1.uiv.p1.Tsealed.context-d1.tau-target-T.ci90.1\endcsname{$-$1.4}
\expandafter\def\csname ppt@rev.desc.d1.uiv.p1.Tsealed.context-d1.tau-target-T.ci90.1\endcsname{$-$1.4}
\expandafter\def\csname val@rev.desc.d1.uiv.p1.Tsealed.context-d1.tau-target-T.ci95-excludes-0\endcsname{1}
\expandafter\def\csname cnt@rev.desc.d1.uiv.p1.Tsealed.context-d1.tau-target-T.ci95-excludes-0\endcsname{1}
\expandafter\def\csname pct@rev.desc.d1.uiv.p1.Tsealed.context-d1.tau-target-T.ci95-excludes-0\endcsname{100.0}
\expandafter\def\csname ppt@rev.desc.d1.uiv.p1.Tsealed.context-d1.tau-target-T.ci95-excludes-0\endcsname{$+$100.0}
\expandafter\def\csname val@rev.desc.d1.uiv.p1.Tsealed.context-d1.tau-target-T.ci95.0\endcsname{-0.111}
\expandafter\def\csname cnt@rev.desc.d1.uiv.p1.Tsealed.context-d1.tau-target-T.ci95.0\endcsname{0}
\expandafter\def\csname pct@rev.desc.d1.uiv.p1.Tsealed.context-d1.tau-target-T.ci95.0\endcsname{$-$11.1}
\expandafter\def\csname ppt@rev.desc.d1.uiv.p1.Tsealed.context-d1.tau-target-T.ci95.0\endcsname{$-$11.1}
\expandafter\def\csname val@rev.desc.d1.uiv.p1.Tsealed.context-d1.tau-target-T.ci95.1\endcsname{-0.007}
\expandafter\def\csname cnt@rev.desc.d1.uiv.p1.Tsealed.context-d1.tau-target-T.ci95.1\endcsname{0}
\expandafter\def\csname pct@rev.desc.d1.uiv.p1.Tsealed.context-d1.tau-target-T.ci95.1\endcsname{$-$0.7}
\expandafter\def\csname ppt@rev.desc.d1.uiv.p1.Tsealed.context-d1.tau-target-T.ci95.1\endcsname{$-$0.7}
\expandafter\def\csname val@rev.desc.d1.uiv.p1.Tsealed.context-d1.tau-target-T.mean\endcsname{-0.056}
\expandafter\def\csname cnt@rev.desc.d1.uiv.p1.Tsealed.context-d1.tau-target-T.mean\endcsname{0}
\expandafter\def\csname pct@rev.desc.d1.uiv.p1.Tsealed.context-d1.tau-target-T.mean\endcsname{$-$5.6}
\expandafter\def\csname ppt@rev.desc.d1.uiv.p1.Tsealed.context-d1.tau-target-T.mean\endcsname{$-$5.6}
\expandafter\def\csname val@rev.desc.d1.uiv.p1.Tsealed.context-d1.tau-target-T.n-apps\endcsname{24}
\expandafter\def\csname cnt@rev.desc.d1.uiv.p1.Tsealed.context-d1.tau-target-T.n-apps\endcsname{24}
\expandafter\def\csname pct@rev.desc.d1.uiv.p1.Tsealed.context-d1.tau-target-T.n-apps\endcsname{2400.0}
\expandafter\def\csname ppt@rev.desc.d1.uiv.p1.Tsealed.context-d1.tau-target-T.n-apps\endcsname{$+$2400.0}
\expandafter\def\csname val@rev.desc.d1.uiv.p1.Tsealed.context-d1.tau-target-T.tost1.ci90-inside\endcsname{0}
\expandafter\def\csname cnt@rev.desc.d1.uiv.p1.Tsealed.context-d1.tau-target-T.tost1.ci90-inside\endcsname{0}
\expandafter\def\csname pct@rev.desc.d1.uiv.p1.Tsealed.context-d1.tau-target-T.tost1.ci90-inside\endcsname{0.0}
\expandafter\def\csname ppt@rev.desc.d1.uiv.p1.Tsealed.context-d1.tau-target-T.tost1.ci90-inside\endcsname{0.0}
\expandafter\def\csname val@rev.desc.d1.uiv.p1.Tsealed.context-d1.tau-target-T.tost1.equivalent\endcsname{0}
\expandafter\def\csname cnt@rev.desc.d1.uiv.p1.Tsealed.context-d1.tau-target-T.tost1.equivalent\endcsname{0}
\expandafter\def\csname pct@rev.desc.d1.uiv.p1.Tsealed.context-d1.tau-target-T.tost1.equivalent\endcsname{0.0}
\expandafter\def\csname ppt@rev.desc.d1.uiv.p1.Tsealed.context-d1.tau-target-T.tost1.equivalent\endcsname{0.0}
\expandafter\def\csname val@rev.desc.d1.uiv.p1.Tsealed.context-d1.tau-target-T.tost1.p\endcsname{0.968}
\expandafter\def\csname cnt@rev.desc.d1.uiv.p1.Tsealed.context-d1.tau-target-T.tost1.p\endcsname{1}
\expandafter\def\csname pct@rev.desc.d1.uiv.p1.Tsealed.context-d1.tau-target-T.tost1.p\endcsname{96.8}
\expandafter\def\csname ppt@rev.desc.d1.uiv.p1.Tsealed.context-d1.tau-target-T.tost1.p\endcsname{$+$96.8}
\expandafter\def\csname val@rev.desc.d1.uiv.p1.Tsealed.context-d1.tau-target-T.tost5.ci90-inside\endcsname{0}
\expandafter\def\csname cnt@rev.desc.d1.uiv.p1.Tsealed.context-d1.tau-target-T.tost5.ci90-inside\endcsname{0}
\expandafter\def\csname pct@rev.desc.d1.uiv.p1.Tsealed.context-d1.tau-target-T.tost5.ci90-inside\endcsname{0.0}
\expandafter\def\csname ppt@rev.desc.d1.uiv.p1.Tsealed.context-d1.tau-target-T.tost5.ci90-inside\endcsname{0.0}
\expandafter\def\csname val@rev.desc.d1.uiv.p1.Tsealed.context-d1.tau-target-T.tost5.equivalent\endcsname{0}
\expandafter\def\csname cnt@rev.desc.d1.uiv.p1.Tsealed.context-d1.tau-target-T.tost5.equivalent\endcsname{0}
\expandafter\def\csname pct@rev.desc.d1.uiv.p1.Tsealed.context-d1.tau-target-T.tost5.equivalent\endcsname{0.0}
\expandafter\def\csname ppt@rev.desc.d1.uiv.p1.Tsealed.context-d1.tau-target-T.tost5.equivalent\endcsname{0.0}
\expandafter\def\csname val@rev.desc.d1.uiv.p1.Tsealed.context-d1.tau-target-T.tost5.p\endcsname{0.516}
\expandafter\def\csname cnt@rev.desc.d1.uiv.p1.Tsealed.context-d1.tau-target-T.tost5.p\endcsname{1}
\expandafter\def\csname pct@rev.desc.d1.uiv.p1.Tsealed.context-d1.tau-target-T.tost5.p\endcsname{51.6}
\expandafter\def\csname ppt@rev.desc.d1.uiv.p1.Tsealed.context-d1.tau-target-T.tost5.p\endcsname{$+$51.6}
\expandafter\def\csname val@rev.desc.d1.uiv.p1.Tsealed.context-d1.zero-estimate\endcsname{0}
\expandafter\def\csname cnt@rev.desc.d1.uiv.p1.Tsealed.context-d1.zero-estimate\endcsname{0}
\expandafter\def\csname pct@rev.desc.d1.uiv.p1.Tsealed.context-d1.zero-estimate\endcsname{0.0}
\expandafter\def\csname ppt@rev.desc.d1.uiv.p1.Tsealed.context-d1.zero-estimate\endcsname{0.0}
\expandafter\def\csname val@rev.desc.d1.uiv.p1.Tsealed.dual-d2.delta-tau-cross.ci90.0\endcsname{-0.118}
\expandafter\def\csname cnt@rev.desc.d1.uiv.p1.Tsealed.dual-d2.delta-tau-cross.ci90.0\endcsname{0}
\expandafter\def\csname pct@rev.desc.d1.uiv.p1.Tsealed.dual-d2.delta-tau-cross.ci90.0\endcsname{$-$11.8}
\expandafter\def\csname ppt@rev.desc.d1.uiv.p1.Tsealed.dual-d2.delta-tau-cross.ci90.0\endcsname{$-$11.8}
\expandafter\def\csname val@rev.desc.d1.uiv.p1.Tsealed.dual-d2.delta-tau-cross.ci90.1\endcsname{0.063}
\expandafter\def\csname cnt@rev.desc.d1.uiv.p1.Tsealed.dual-d2.delta-tau-cross.ci90.1\endcsname{0}
\expandafter\def\csname pct@rev.desc.d1.uiv.p1.Tsealed.dual-d2.delta-tau-cross.ci90.1\endcsname{6.3}
\expandafter\def\csname ppt@rev.desc.d1.uiv.p1.Tsealed.dual-d2.delta-tau-cross.ci90.1\endcsname{$+$6.3}
\expandafter\def\csname val@rev.desc.d1.uiv.p1.Tsealed.dual-d2.delta-tau-cross.ci95-excludes-0\endcsname{0}
\expandafter\def\csname cnt@rev.desc.d1.uiv.p1.Tsealed.dual-d2.delta-tau-cross.ci95-excludes-0\endcsname{0}
\expandafter\def\csname pct@rev.desc.d1.uiv.p1.Tsealed.dual-d2.delta-tau-cross.ci95-excludes-0\endcsname{0.0}
\expandafter\def\csname ppt@rev.desc.d1.uiv.p1.Tsealed.dual-d2.delta-tau-cross.ci95-excludes-0\endcsname{0.0}
\expandafter\def\csname val@rev.desc.d1.uiv.p1.Tsealed.dual-d2.delta-tau-cross.ci95.0\endcsname{-0.139}
\expandafter\def\csname cnt@rev.desc.d1.uiv.p1.Tsealed.dual-d2.delta-tau-cross.ci95.0\endcsname{0}
\expandafter\def\csname pct@rev.desc.d1.uiv.p1.Tsealed.dual-d2.delta-tau-cross.ci95.0\endcsname{$-$13.9}
\expandafter\def\csname ppt@rev.desc.d1.uiv.p1.Tsealed.dual-d2.delta-tau-cross.ci95.0\endcsname{$-$13.9}
\expandafter\def\csname val@rev.desc.d1.uiv.p1.Tsealed.dual-d2.delta-tau-cross.ci95.1\endcsname{0.076}
\expandafter\def\csname cnt@rev.desc.d1.uiv.p1.Tsealed.dual-d2.delta-tau-cross.ci95.1\endcsname{0}
\expandafter\def\csname pct@rev.desc.d1.uiv.p1.Tsealed.dual-d2.delta-tau-cross.ci95.1\endcsname{7.6}
\expandafter\def\csname ppt@rev.desc.d1.uiv.p1.Tsealed.dual-d2.delta-tau-cross.ci95.1\endcsname{$+$7.6}
\expandafter\def\csname val@rev.desc.d1.uiv.p1.Tsealed.dual-d2.delta-tau-cross.mean\endcsname{-0.021}
\expandafter\def\csname cnt@rev.desc.d1.uiv.p1.Tsealed.dual-d2.delta-tau-cross.mean\endcsname{0}
\expandafter\def\csname pct@rev.desc.d1.uiv.p1.Tsealed.dual-d2.delta-tau-cross.mean\endcsname{$-$2.1}
\expandafter\def\csname ppt@rev.desc.d1.uiv.p1.Tsealed.dual-d2.delta-tau-cross.mean\endcsname{$-$2.1}
\expandafter\def\csname val@rev.desc.d1.uiv.p1.Tsealed.dual-d2.delta-tau-cross.n-apps.0\endcsname{24}
\expandafter\def\csname cnt@rev.desc.d1.uiv.p1.Tsealed.dual-d2.delta-tau-cross.n-apps.0\endcsname{24}
\expandafter\def\csname pct@rev.desc.d1.uiv.p1.Tsealed.dual-d2.delta-tau-cross.n-apps.0\endcsname{2400.0}
\expandafter\def\csname ppt@rev.desc.d1.uiv.p1.Tsealed.dual-d2.delta-tau-cross.n-apps.0\endcsname{$+$2400.0}
\expandafter\def\csname val@rev.desc.d1.uiv.p1.Tsealed.dual-d2.delta-tau-cross.n-apps.1\endcsname{12}
\expandafter\def\csname cnt@rev.desc.d1.uiv.p1.Tsealed.dual-d2.delta-tau-cross.n-apps.1\endcsname{12}
\expandafter\def\csname pct@rev.desc.d1.uiv.p1.Tsealed.dual-d2.delta-tau-cross.n-apps.1\endcsname{1200.0}
\expandafter\def\csname ppt@rev.desc.d1.uiv.p1.Tsealed.dual-d2.delta-tau-cross.n-apps.1\endcsname{$+$1200.0}
\expandafter\def\csname val@rev.desc.d1.uiv.p1.Tsealed.dual-d2.delta-tau-cross.tost1.ci90-inside\endcsname{0}
\expandafter\def\csname cnt@rev.desc.d1.uiv.p1.Tsealed.dual-d2.delta-tau-cross.tost1.ci90-inside\endcsname{0}
\expandafter\def\csname pct@rev.desc.d1.uiv.p1.Tsealed.dual-d2.delta-tau-cross.tost1.ci90-inside\endcsname{0.0}
\expandafter\def\csname ppt@rev.desc.d1.uiv.p1.Tsealed.dual-d2.delta-tau-cross.tost1.ci90-inside\endcsname{0.0}
\expandafter\def\csname val@rev.desc.d1.uiv.p1.Tsealed.dual-d2.delta-tau-cross.tost1.equivalent\endcsname{0}
\expandafter\def\csname cnt@rev.desc.d1.uiv.p1.Tsealed.dual-d2.delta-tau-cross.tost1.equivalent\endcsname{0}
\expandafter\def\csname pct@rev.desc.d1.uiv.p1.Tsealed.dual-d2.delta-tau-cross.tost1.equivalent\endcsname{0.0}
\expandafter\def\csname ppt@rev.desc.d1.uiv.p1.Tsealed.dual-d2.delta-tau-cross.tost1.equivalent\endcsname{0.0}
\expandafter\def\csname val@rev.desc.d1.uiv.p1.Tsealed.dual-d2.delta-tau-cross.tost1.p\endcsname{0.578}
\expandafter\def\csname cnt@rev.desc.d1.uiv.p1.Tsealed.dual-d2.delta-tau-cross.tost1.p\endcsname{1}
\expandafter\def\csname pct@rev.desc.d1.uiv.p1.Tsealed.dual-d2.delta-tau-cross.tost1.p\endcsname{57.8}
\expandafter\def\csname ppt@rev.desc.d1.uiv.p1.Tsealed.dual-d2.delta-tau-cross.tost1.p\endcsname{$+$57.8}
\expandafter\def\csname val@rev.desc.d1.uiv.p1.Tsealed.dual-d2.delta-tau-cross.tost5.ci90-inside\endcsname{0}
\expandafter\def\csname cnt@rev.desc.d1.uiv.p1.Tsealed.dual-d2.delta-tau-cross.tost5.ci90-inside\endcsname{0}
\expandafter\def\csname pct@rev.desc.d1.uiv.p1.Tsealed.dual-d2.delta-tau-cross.tost5.ci90-inside\endcsname{0.0}
\expandafter\def\csname ppt@rev.desc.d1.uiv.p1.Tsealed.dual-d2.delta-tau-cross.tost5.ci90-inside\endcsname{0.0}
\expandafter\def\csname val@rev.desc.d1.uiv.p1.Tsealed.dual-d2.delta-tau-cross.tost5.equivalent\endcsname{0}
\expandafter\def\csname cnt@rev.desc.d1.uiv.p1.Tsealed.dual-d2.delta-tau-cross.tost5.equivalent\endcsname{0}
\expandafter\def\csname pct@rev.desc.d1.uiv.p1.Tsealed.dual-d2.delta-tau-cross.tost5.equivalent\endcsname{0.0}
\expandafter\def\csname ppt@rev.desc.d1.uiv.p1.Tsealed.dual-d2.delta-tau-cross.tost5.equivalent\endcsname{0.0}
\expandafter\def\csname val@rev.desc.d1.uiv.p1.Tsealed.dual-d2.delta-tau-cross.tost5.p\endcsname{0.285}
\expandafter\def\csname cnt@rev.desc.d1.uiv.p1.Tsealed.dual-d2.delta-tau-cross.tost5.p\endcsname{0}
\expandafter\def\csname pct@rev.desc.d1.uiv.p1.Tsealed.dual-d2.delta-tau-cross.tost5.p\endcsname{28.5}
\expandafter\def\csname ppt@rev.desc.d1.uiv.p1.Tsealed.dual-d2.delta-tau-cross.tost5.p\endcsname{$+$28.5}
\expandafter\def\csname val@rev.desc.d1.uiv.p1.Tsealed.dual-d2.material-flip\endcsname{0}
\expandafter\def\csname cnt@rev.desc.d1.uiv.p1.Tsealed.dual-d2.material-flip\endcsname{0}
\expandafter\def\csname pct@rev.desc.d1.uiv.p1.Tsealed.dual-d2.material-flip\endcsname{0.0}
\expandafter\def\csname ppt@rev.desc.d1.uiv.p1.Tsealed.dual-d2.material-flip\endcsname{0.0}
\expandafter\def\csname val@rev.desc.d1.uiv.p1.Tsealed.dual-d2.n-tasks\endcsname{144}
\expandafter\def\csname cnt@rev.desc.d1.uiv.p1.Tsealed.dual-d2.n-tasks\endcsname{144}
\expandafter\def\csname pct@rev.desc.d1.uiv.p1.Tsealed.dual-d2.n-tasks\endcsname{14400.0}
\expandafter\def\csname ppt@rev.desc.d1.uiv.p1.Tsealed.dual-d2.n-tasks\endcsname{$+$14400.0}
\expandafter\def\csname val@rev.desc.d1.uiv.p1.Tsealed.dual-d2.n-tasks-source-C\endcsname{96}
\expandafter\def\csname cnt@rev.desc.d1.uiv.p1.Tsealed.dual-d2.n-tasks-source-C\endcsname{96}
\expandafter\def\csname pct@rev.desc.d1.uiv.p1.Tsealed.dual-d2.n-tasks-source-C\endcsname{9600.0}
\expandafter\def\csname ppt@rev.desc.d1.uiv.p1.Tsealed.dual-d2.n-tasks-source-C\endcsname{$+$9600.0}
\expandafter\def\csname val@rev.desc.d1.uiv.p1.Tsealed.dual-d2.paper-tau-T\endcsname{-0.104}
\expandafter\def\csname cnt@rev.desc.d1.uiv.p1.Tsealed.dual-d2.paper-tau-T\endcsname{0}
\expandafter\def\csname pct@rev.desc.d1.uiv.p1.Tsealed.dual-d2.paper-tau-T\endcsname{$-$10.4}
\expandafter\def\csname ppt@rev.desc.d1.uiv.p1.Tsealed.dual-d2.paper-tau-T\endcsname{$-$10.4}
\expandafter\def\csname val@rev.desc.d1.uiv.p1.Tsealed.dual-d2.reproduces\endcsname{1}
\expandafter\def\csname cnt@rev.desc.d1.uiv.p1.Tsealed.dual-d2.reproduces\endcsname{1}
\expandafter\def\csname pct@rev.desc.d1.uiv.p1.Tsealed.dual-d2.reproduces\endcsname{100.0}
\expandafter\def\csname ppt@rev.desc.d1.uiv.p1.Tsealed.dual-d2.reproduces\endcsname{$+$100.0}
\expandafter\def\csname val@rev.desc.d1.uiv.p1.Tsealed.dual-d2.reproduces-ci\endcsname{1}
\expandafter\def\csname cnt@rev.desc.d1.uiv.p1.Tsealed.dual-d2.reproduces-ci\endcsname{1}
\expandafter\def\csname pct@rev.desc.d1.uiv.p1.Tsealed.dual-d2.reproduces-ci\endcsname{100.0}
\expandafter\def\csname ppt@rev.desc.d1.uiv.p1.Tsealed.dual-d2.reproduces-ci\endcsname{$+$100.0}
\expandafter\def\csname val@rev.desc.d1.uiv.p1.Tsealed.dual-d2.sign-differs\endcsname{0}
\expandafter\def\csname cnt@rev.desc.d1.uiv.p1.Tsealed.dual-d2.sign-differs\endcsname{0}
\expandafter\def\csname pct@rev.desc.d1.uiv.p1.Tsealed.dual-d2.sign-differs\endcsname{0.0}
\expandafter\def\csname ppt@rev.desc.d1.uiv.p1.Tsealed.dual-d2.sign-differs\endcsname{0.0}
\expandafter\def\csname val@rev.desc.d1.uiv.p1.Tsealed.dual-d2.tau-source-C.ci90.0\endcsname{-0.156}
\expandafter\def\csname cnt@rev.desc.d1.uiv.p1.Tsealed.dual-d2.tau-source-C.ci90.0\endcsname{0}
\expandafter\def\csname pct@rev.desc.d1.uiv.p1.Tsealed.dual-d2.tau-source-C.ci90.0\endcsname{$-$15.6}
\expandafter\def\csname ppt@rev.desc.d1.uiv.p1.Tsealed.dual-d2.tau-source-C.ci90.0\endcsname{$-$15.6}
\expandafter\def\csname val@rev.desc.d1.uiv.p1.Tsealed.dual-d2.tau-source-C.ci90.1\endcsname{0.000}
\expandafter\def\csname cnt@rev.desc.d1.uiv.p1.Tsealed.dual-d2.tau-source-C.ci90.1\endcsname{0}
\expandafter\def\csname pct@rev.desc.d1.uiv.p1.Tsealed.dual-d2.tau-source-C.ci90.1\endcsname{0.0}
\expandafter\def\csname ppt@rev.desc.d1.uiv.p1.Tsealed.dual-d2.tau-source-C.ci90.1\endcsname{0.0}
\expandafter\def\csname val@rev.desc.d1.uiv.p1.Tsealed.dual-d2.tau-source-C.ci95-excludes-0\endcsname{0}
\expandafter\def\csname cnt@rev.desc.d1.uiv.p1.Tsealed.dual-d2.tau-source-C.ci95-excludes-0\endcsname{0}
\expandafter\def\csname pct@rev.desc.d1.uiv.p1.Tsealed.dual-d2.tau-source-C.ci95-excludes-0\endcsname{0.0}
\expandafter\def\csname ppt@rev.desc.d1.uiv.p1.Tsealed.dual-d2.tau-source-C.ci95-excludes-0\endcsname{0.0}
\expandafter\def\csname val@rev.desc.d1.uiv.p1.Tsealed.dual-d2.tau-source-C.ci95.0\endcsname{-0.167}
\expandafter\def\csname cnt@rev.desc.d1.uiv.p1.Tsealed.dual-d2.tau-source-C.ci95.0\endcsname{0}
\expandafter\def\csname pct@rev.desc.d1.uiv.p1.Tsealed.dual-d2.tau-source-C.ci95.0\endcsname{$-$16.7}
\expandafter\def\csname ppt@rev.desc.d1.uiv.p1.Tsealed.dual-d2.tau-source-C.ci95.0\endcsname{$-$16.7}
\expandafter\def\csname val@rev.desc.d1.uiv.p1.Tsealed.dual-d2.tau-source-C.ci95.1\endcsname{0.021}
\expandafter\def\csname cnt@rev.desc.d1.uiv.p1.Tsealed.dual-d2.tau-source-C.ci95.1\endcsname{0}
\expandafter\def\csname pct@rev.desc.d1.uiv.p1.Tsealed.dual-d2.tau-source-C.ci95.1\endcsname{2.1}
\expandafter\def\csname ppt@rev.desc.d1.uiv.p1.Tsealed.dual-d2.tau-source-C.ci95.1\endcsname{$+$2.1}
\expandafter\def\csname val@rev.desc.d1.uiv.p1.Tsealed.dual-d2.tau-source-C.mean\endcsname{-0.083}
\expandafter\def\csname cnt@rev.desc.d1.uiv.p1.Tsealed.dual-d2.tau-source-C.mean\endcsname{0}
\expandafter\def\csname pct@rev.desc.d1.uiv.p1.Tsealed.dual-d2.tau-source-C.mean\endcsname{$-$8.3}
\expandafter\def\csname ppt@rev.desc.d1.uiv.p1.Tsealed.dual-d2.tau-source-C.mean\endcsname{$-$8.3}
\expandafter\def\csname val@rev.desc.d1.uiv.p1.Tsealed.dual-d2.tau-source-C.n-apps\endcsname{12}
\expandafter\def\csname cnt@rev.desc.d1.uiv.p1.Tsealed.dual-d2.tau-source-C.n-apps\endcsname{12}
\expandafter\def\csname pct@rev.desc.d1.uiv.p1.Tsealed.dual-d2.tau-source-C.n-apps\endcsname{1200.0}
\expandafter\def\csname ppt@rev.desc.d1.uiv.p1.Tsealed.dual-d2.tau-source-C.n-apps\endcsname{$+$1200.0}
\expandafter\def\csname val@rev.desc.d1.uiv.p1.Tsealed.dual-d2.tau-source-C.tost1.ci90-inside\endcsname{0}
\expandafter\def\csname cnt@rev.desc.d1.uiv.p1.Tsealed.dual-d2.tau-source-C.tost1.ci90-inside\endcsname{0}
\expandafter\def\csname pct@rev.desc.d1.uiv.p1.Tsealed.dual-d2.tau-source-C.tost1.ci90-inside\endcsname{0.0}
\expandafter\def\csname ppt@rev.desc.d1.uiv.p1.Tsealed.dual-d2.tau-source-C.tost1.ci90-inside\endcsname{0.0}
\expandafter\def\csname val@rev.desc.d1.uiv.p1.Tsealed.dual-d2.tau-source-C.tost1.equivalent\endcsname{0}
\expandafter\def\csname cnt@rev.desc.d1.uiv.p1.Tsealed.dual-d2.tau-source-C.tost1.equivalent\endcsname{0}
\expandafter\def\csname pct@rev.desc.d1.uiv.p1.Tsealed.dual-d2.tau-source-C.tost1.equivalent\endcsname{0.0}
\expandafter\def\csname ppt@rev.desc.d1.uiv.p1.Tsealed.dual-d2.tau-source-C.tost1.equivalent\endcsname{0.0}
\expandafter\def\csname val@rev.desc.d1.uiv.p1.Tsealed.dual-d2.tau-source-C.tost1.p\endcsname{0.946}
\expandafter\def\csname cnt@rev.desc.d1.uiv.p1.Tsealed.dual-d2.tau-source-C.tost1.p\endcsname{1}
\expandafter\def\csname pct@rev.desc.d1.uiv.p1.Tsealed.dual-d2.tau-source-C.tost1.p\endcsname{94.6}
\expandafter\def\csname ppt@rev.desc.d1.uiv.p1.Tsealed.dual-d2.tau-source-C.tost1.p\endcsname{$+$94.6}
\expandafter\def\csname val@rev.desc.d1.uiv.p1.Tsealed.dual-d2.tau-source-C.tost5.ci90-inside\endcsname{0}
\expandafter\def\csname cnt@rev.desc.d1.uiv.p1.Tsealed.dual-d2.tau-source-C.tost5.ci90-inside\endcsname{0}
\expandafter\def\csname pct@rev.desc.d1.uiv.p1.Tsealed.dual-d2.tau-source-C.tost5.ci90-inside\endcsname{0.0}
\expandafter\def\csname ppt@rev.desc.d1.uiv.p1.Tsealed.dual-d2.tau-source-C.tost5.ci90-inside\endcsname{0.0}
\expandafter\def\csname val@rev.desc.d1.uiv.p1.Tsealed.dual-d2.tau-source-C.tost5.equivalent\endcsname{0}
\expandafter\def\csname cnt@rev.desc.d1.uiv.p1.Tsealed.dual-d2.tau-source-C.tost5.equivalent\endcsname{0}
\expandafter\def\csname pct@rev.desc.d1.uiv.p1.Tsealed.dual-d2.tau-source-C.tost5.equivalent\endcsname{0.0}
\expandafter\def\csname ppt@rev.desc.d1.uiv.p1.Tsealed.dual-d2.tau-source-C.tost5.equivalent\endcsname{0.0}
\expandafter\def\csname val@rev.desc.d1.uiv.p1.Tsealed.dual-d2.tau-source-C.tost5.p\endcsname{0.790}
\expandafter\def\csname cnt@rev.desc.d1.uiv.p1.Tsealed.dual-d2.tau-source-C.tost5.p\endcsname{1}
\expandafter\def\csname pct@rev.desc.d1.uiv.p1.Tsealed.dual-d2.tau-source-C.tost5.p\endcsname{79.0}
\expandafter\def\csname ppt@rev.desc.d1.uiv.p1.Tsealed.dual-d2.tau-source-C.tost5.p\endcsname{$+$79.0}
\expandafter\def\csname val@rev.desc.d1.uiv.p1.Tsealed.dual-d2.tau-target-T.ci90.0\endcsname{-0.146}
\expandafter\def\csname cnt@rev.desc.d1.uiv.p1.Tsealed.dual-d2.tau-target-T.ci90.0\endcsname{0}
\expandafter\def\csname pct@rev.desc.d1.uiv.p1.Tsealed.dual-d2.tau-target-T.ci90.0\endcsname{$-$14.6}
\expandafter\def\csname ppt@rev.desc.d1.uiv.p1.Tsealed.dual-d2.tau-target-T.ci90.0\endcsname{$-$14.6}
\expandafter\def\csname val@rev.desc.d1.uiv.p1.Tsealed.dual-d2.tau-target-T.ci90.1\endcsname{-0.062}
\expandafter\def\csname cnt@rev.desc.d1.uiv.p1.Tsealed.dual-d2.tau-target-T.ci90.1\endcsname{0}
\expandafter\def\csname pct@rev.desc.d1.uiv.p1.Tsealed.dual-d2.tau-target-T.ci90.1\endcsname{$-$6.2}
\expandafter\def\csname ppt@rev.desc.d1.uiv.p1.Tsealed.dual-d2.tau-target-T.ci90.1\endcsname{$-$6.2}
\expandafter\def\csname val@rev.desc.d1.uiv.p1.Tsealed.dual-d2.tau-target-T.ci95-excludes-0\endcsname{1}
\expandafter\def\csname cnt@rev.desc.d1.uiv.p1.Tsealed.dual-d2.tau-target-T.ci95-excludes-0\endcsname{1}
\expandafter\def\csname pct@rev.desc.d1.uiv.p1.Tsealed.dual-d2.tau-target-T.ci95-excludes-0\endcsname{100.0}
\expandafter\def\csname ppt@rev.desc.d1.uiv.p1.Tsealed.dual-d2.tau-target-T.ci95-excludes-0\endcsname{$+$100.0}
\expandafter\def\csname val@rev.desc.d1.uiv.p1.Tsealed.dual-d2.tau-target-T.ci95.0\endcsname{-0.160}
\expandafter\def\csname cnt@rev.desc.d1.uiv.p1.Tsealed.dual-d2.tau-target-T.ci95.0\endcsname{0}
\expandafter\def\csname pct@rev.desc.d1.uiv.p1.Tsealed.dual-d2.tau-target-T.ci95.0\endcsname{$-$16.0}
\expandafter\def\csname ppt@rev.desc.d1.uiv.p1.Tsealed.dual-d2.tau-target-T.ci95.0\endcsname{$-$16.0}
\expandafter\def\csname val@rev.desc.d1.uiv.p1.Tsealed.dual-d2.tau-target-T.ci95.1\endcsname{-0.049}
\expandafter\def\csname cnt@rev.desc.d1.uiv.p1.Tsealed.dual-d2.tau-target-T.ci95.1\endcsname{0}
\expandafter\def\csname pct@rev.desc.d1.uiv.p1.Tsealed.dual-d2.tau-target-T.ci95.1\endcsname{$-$4.9}
\expandafter\def\csname ppt@rev.desc.d1.uiv.p1.Tsealed.dual-d2.tau-target-T.ci95.1\endcsname{$-$4.9}
\expandafter\def\csname val@rev.desc.d1.uiv.p1.Tsealed.dual-d2.tau-target-T.mean\endcsname{-0.104}
\expandafter\def\csname cnt@rev.desc.d1.uiv.p1.Tsealed.dual-d2.tau-target-T.mean\endcsname{0}
\expandafter\def\csname pct@rev.desc.d1.uiv.p1.Tsealed.dual-d2.tau-target-T.mean\endcsname{$-$10.4}
\expandafter\def\csname ppt@rev.desc.d1.uiv.p1.Tsealed.dual-d2.tau-target-T.mean\endcsname{$-$10.4}
\expandafter\def\csname val@rev.desc.d1.uiv.p1.Tsealed.dual-d2.tau-target-T.n-apps\endcsname{24}
\expandafter\def\csname cnt@rev.desc.d1.uiv.p1.Tsealed.dual-d2.tau-target-T.n-apps\endcsname{24}
\expandafter\def\csname pct@rev.desc.d1.uiv.p1.Tsealed.dual-d2.tau-target-T.n-apps\endcsname{2400.0}
\expandafter\def\csname ppt@rev.desc.d1.uiv.p1.Tsealed.dual-d2.tau-target-T.n-apps\endcsname{$+$2400.0}
\expandafter\def\csname val@rev.desc.d1.uiv.p1.Tsealed.dual-d2.tau-target-T.tost1.ci90-inside\endcsname{0}
\expandafter\def\csname cnt@rev.desc.d1.uiv.p1.Tsealed.dual-d2.tau-target-T.tost1.ci90-inside\endcsname{0}
\expandafter\def\csname pct@rev.desc.d1.uiv.p1.Tsealed.dual-d2.tau-target-T.tost1.ci90-inside\endcsname{0.0}
\expandafter\def\csname ppt@rev.desc.d1.uiv.p1.Tsealed.dual-d2.tau-target-T.tost1.ci90-inside\endcsname{0.0}
\expandafter\def\csname val@rev.desc.d1.uiv.p1.Tsealed.dual-d2.tau-target-T.tost1.equivalent\endcsname{0}
\expandafter\def\csname cnt@rev.desc.d1.uiv.p1.Tsealed.dual-d2.tau-target-T.tost1.equivalent\endcsname{0}
\expandafter\def\csname pct@rev.desc.d1.uiv.p1.Tsealed.dual-d2.tau-target-T.tost1.equivalent\endcsname{0.0}
\expandafter\def\csname ppt@rev.desc.d1.uiv.p1.Tsealed.dual-d2.tau-target-T.tost1.equivalent\endcsname{0.0}
\expandafter\def\csname val@rev.desc.d1.uiv.p1.Tsealed.dual-d2.tau-target-T.tost1.p\endcsname{0.999}
\expandafter\def\csname cnt@rev.desc.d1.uiv.p1.Tsealed.dual-d2.tau-target-T.tost1.p\endcsname{1}
\expandafter\def\csname pct@rev.desc.d1.uiv.p1.Tsealed.dual-d2.tau-target-T.tost1.p\endcsname{99.9}
\expandafter\def\csname ppt@rev.desc.d1.uiv.p1.Tsealed.dual-d2.tau-target-T.tost1.p\endcsname{$+$99.9}
\expandafter\def\csname val@rev.desc.d1.uiv.p1.Tsealed.dual-d2.tau-target-T.tost5.ci90-inside\endcsname{0}
\expandafter\def\csname cnt@rev.desc.d1.uiv.p1.Tsealed.dual-d2.tau-target-T.tost5.ci90-inside\endcsname{0}
\expandafter\def\csname pct@rev.desc.d1.uiv.p1.Tsealed.dual-d2.tau-target-T.tost5.ci90-inside\endcsname{0.0}
\expandafter\def\csname ppt@rev.desc.d1.uiv.p1.Tsealed.dual-d2.tau-target-T.tost5.ci90-inside\endcsname{0.0}
\expandafter\def\csname val@rev.desc.d1.uiv.p1.Tsealed.dual-d2.tau-target-T.tost5.equivalent\endcsname{0}
\expandafter\def\csname cnt@rev.desc.d1.uiv.p1.Tsealed.dual-d2.tau-target-T.tost5.equivalent\endcsname{0}
\expandafter\def\csname pct@rev.desc.d1.uiv.p1.Tsealed.dual-d2.tau-target-T.tost5.equivalent\endcsname{0.0}
\expandafter\def\csname ppt@rev.desc.d1.uiv.p1.Tsealed.dual-d2.tau-target-T.tost5.equivalent\endcsname{0.0}
\expandafter\def\csname val@rev.desc.d1.uiv.p1.Tsealed.dual-d2.tau-target-T.tost5.p\endcsname{0.973}
\expandafter\def\csname cnt@rev.desc.d1.uiv.p1.Tsealed.dual-d2.tau-target-T.tost5.p\endcsname{1}
\expandafter\def\csname pct@rev.desc.d1.uiv.p1.Tsealed.dual-d2.tau-target-T.tost5.p\endcsname{97.3}
\expandafter\def\csname ppt@rev.desc.d1.uiv.p1.Tsealed.dual-d2.tau-target-T.tost5.p\endcsname{$+$97.3}
\expandafter\def\csname val@rev.desc.d1.uiv.p1.Tsealed.dual-d2.zero-estimate\endcsname{0}
\expandafter\def\csname cnt@rev.desc.d1.uiv.p1.Tsealed.dual-d2.zero-estimate\endcsname{0}
\expandafter\def\csname pct@rev.desc.d1.uiv.p1.Tsealed.dual-d2.zero-estimate\endcsname{0.0}
\expandafter\def\csname ppt@rev.desc.d1.uiv.p1.Tsealed.dual-d2.zero-estimate\endcsname{0.0}
\expandafter\def\csname val@rev.desc.d1.uiv.p1.Tsealed.grid2-d3.delta-tau-cross.ci90.0\endcsname{-0.014}
\expandafter\def\csname cnt@rev.desc.d1.uiv.p1.Tsealed.grid2-d3.delta-tau-cross.ci90.0\endcsname{0}
\expandafter\def\csname pct@rev.desc.d1.uiv.p1.Tsealed.grid2-d3.delta-tau-cross.ci90.0\endcsname{$-$1.4}
\expandafter\def\csname ppt@rev.desc.d1.uiv.p1.Tsealed.grid2-d3.delta-tau-cross.ci90.0\endcsname{$-$1.4}
\expandafter\def\csname val@rev.desc.d1.uiv.p1.Tsealed.grid2-d3.delta-tau-cross.ci90.1\endcsname{0.153}
\expandafter\def\csname cnt@rev.desc.d1.uiv.p1.Tsealed.grid2-d3.delta-tau-cross.ci90.1\endcsname{0}
\expandafter\def\csname pct@rev.desc.d1.uiv.p1.Tsealed.grid2-d3.delta-tau-cross.ci90.1\endcsname{15.3}
\expandafter\def\csname ppt@rev.desc.d1.uiv.p1.Tsealed.grid2-d3.delta-tau-cross.ci90.1\endcsname{$+$15.3}
\expandafter\def\csname val@rev.desc.d1.uiv.p1.Tsealed.grid2-d3.delta-tau-cross.ci95-excludes-0\endcsname{0}
\expandafter\def\csname cnt@rev.desc.d1.uiv.p1.Tsealed.grid2-d3.delta-tau-cross.ci95-excludes-0\endcsname{0}
\expandafter\def\csname pct@rev.desc.d1.uiv.p1.Tsealed.grid2-d3.delta-tau-cross.ci95-excludes-0\endcsname{0.0}
\expandafter\def\csname ppt@rev.desc.d1.uiv.p1.Tsealed.grid2-d3.delta-tau-cross.ci95-excludes-0\endcsname{0.0}
\expandafter\def\csname val@rev.desc.d1.uiv.p1.Tsealed.grid2-d3.delta-tau-cross.ci95.0\endcsname{-0.028}
\expandafter\def\csname cnt@rev.desc.d1.uiv.p1.Tsealed.grid2-d3.delta-tau-cross.ci95.0\endcsname{0}
\expandafter\def\csname pct@rev.desc.d1.uiv.p1.Tsealed.grid2-d3.delta-tau-cross.ci95.0\endcsname{$-$2.8}
\expandafter\def\csname ppt@rev.desc.d1.uiv.p1.Tsealed.grid2-d3.delta-tau-cross.ci95.0\endcsname{$-$2.8}
\expandafter\def\csname val@rev.desc.d1.uiv.p1.Tsealed.grid2-d3.delta-tau-cross.ci95.1\endcsname{0.174}
\expandafter\def\csname cnt@rev.desc.d1.uiv.p1.Tsealed.grid2-d3.delta-tau-cross.ci95.1\endcsname{0}
\expandafter\def\csname pct@rev.desc.d1.uiv.p1.Tsealed.grid2-d3.delta-tau-cross.ci95.1\endcsname{17.4}
\expandafter\def\csname ppt@rev.desc.d1.uiv.p1.Tsealed.grid2-d3.delta-tau-cross.ci95.1\endcsname{$+$17.4}
\expandafter\def\csname val@rev.desc.d1.uiv.p1.Tsealed.grid2-d3.delta-tau-cross.mean\endcsname{0.069}
\expandafter\def\csname cnt@rev.desc.d1.uiv.p1.Tsealed.grid2-d3.delta-tau-cross.mean\endcsname{0}
\expandafter\def\csname pct@rev.desc.d1.uiv.p1.Tsealed.grid2-d3.delta-tau-cross.mean\endcsname{6.9}
\expandafter\def\csname ppt@rev.desc.d1.uiv.p1.Tsealed.grid2-d3.delta-tau-cross.mean\endcsname{$+$6.9}
\expandafter\def\csname val@rev.desc.d1.uiv.p1.Tsealed.grid2-d3.delta-tau-cross.n-apps.0\endcsname{24}
\expandafter\def\csname cnt@rev.desc.d1.uiv.p1.Tsealed.grid2-d3.delta-tau-cross.n-apps.0\endcsname{24}
\expandafter\def\csname pct@rev.desc.d1.uiv.p1.Tsealed.grid2-d3.delta-tau-cross.n-apps.0\endcsname{2400.0}
\expandafter\def\csname ppt@rev.desc.d1.uiv.p1.Tsealed.grid2-d3.delta-tau-cross.n-apps.0\endcsname{$+$2400.0}
\expandafter\def\csname val@rev.desc.d1.uiv.p1.Tsealed.grid2-d3.delta-tau-cross.n-apps.1\endcsname{12}
\expandafter\def\csname cnt@rev.desc.d1.uiv.p1.Tsealed.grid2-d3.delta-tau-cross.n-apps.1\endcsname{12}
\expandafter\def\csname pct@rev.desc.d1.uiv.p1.Tsealed.grid2-d3.delta-tau-cross.n-apps.1\endcsname{1200.0}
\expandafter\def\csname ppt@rev.desc.d1.uiv.p1.Tsealed.grid2-d3.delta-tau-cross.n-apps.1\endcsname{$+$1200.0}
\expandafter\def\csname val@rev.desc.d1.uiv.p1.Tsealed.grid2-d3.delta-tau-cross.tost1.ci90-inside\endcsname{0}
\expandafter\def\csname cnt@rev.desc.d1.uiv.p1.Tsealed.grid2-d3.delta-tau-cross.tost1.ci90-inside\endcsname{0}
\expandafter\def\csname pct@rev.desc.d1.uiv.p1.Tsealed.grid2-d3.delta-tau-cross.tost1.ci90-inside\endcsname{0.0}
\expandafter\def\csname ppt@rev.desc.d1.uiv.p1.Tsealed.grid2-d3.delta-tau-cross.tost1.ci90-inside\endcsname{0.0}
\expandafter\def\csname val@rev.desc.d1.uiv.p1.Tsealed.grid2-d3.delta-tau-cross.tost1.equivalent\endcsname{0}
\expandafter\def\csname cnt@rev.desc.d1.uiv.p1.Tsealed.grid2-d3.delta-tau-cross.tost1.equivalent\endcsname{0}
\expandafter\def\csname pct@rev.desc.d1.uiv.p1.Tsealed.grid2-d3.delta-tau-cross.tost1.equivalent\endcsname{0.0}
\expandafter\def\csname ppt@rev.desc.d1.uiv.p1.Tsealed.grid2-d3.delta-tau-cross.tost1.equivalent\endcsname{0.0}
\expandafter\def\csname val@rev.desc.d1.uiv.p1.Tsealed.grid2-d3.delta-tau-cross.tost1.p\endcsname{0.886}
\expandafter\def\csname cnt@rev.desc.d1.uiv.p1.Tsealed.grid2-d3.delta-tau-cross.tost1.p\endcsname{1}
\expandafter\def\csname pct@rev.desc.d1.uiv.p1.Tsealed.grid2-d3.delta-tau-cross.tost1.p\endcsname{88.6}
\expandafter\def\csname ppt@rev.desc.d1.uiv.p1.Tsealed.grid2-d3.delta-tau-cross.tost1.p\endcsname{$+$88.6}
\expandafter\def\csname val@rev.desc.d1.uiv.p1.Tsealed.grid2-d3.delta-tau-cross.tost5.ci90-inside\endcsname{0}
\expandafter\def\csname cnt@rev.desc.d1.uiv.p1.Tsealed.grid2-d3.delta-tau-cross.tost5.ci90-inside\endcsname{0}
\expandafter\def\csname pct@rev.desc.d1.uiv.p1.Tsealed.grid2-d3.delta-tau-cross.tost5.ci90-inside\endcsname{0.0}
\expandafter\def\csname ppt@rev.desc.d1.uiv.p1.Tsealed.grid2-d3.delta-tau-cross.tost5.ci90-inside\endcsname{0.0}
\expandafter\def\csname val@rev.desc.d1.uiv.p1.Tsealed.grid2-d3.delta-tau-cross.tost5.equivalent\endcsname{0}
\expandafter\def\csname cnt@rev.desc.d1.uiv.p1.Tsealed.grid2-d3.delta-tau-cross.tost5.equivalent\endcsname{0}
\expandafter\def\csname pct@rev.desc.d1.uiv.p1.Tsealed.grid2-d3.delta-tau-cross.tost5.equivalent\endcsname{0.0}
\expandafter\def\csname ppt@rev.desc.d1.uiv.p1.Tsealed.grid2-d3.delta-tau-cross.tost5.equivalent\endcsname{0.0}
\expandafter\def\csname val@rev.desc.d1.uiv.p1.Tsealed.grid2-d3.delta-tau-cross.tost5.p\endcsname{0.650}
\expandafter\def\csname cnt@rev.desc.d1.uiv.p1.Tsealed.grid2-d3.delta-tau-cross.tost5.p\endcsname{1}
\expandafter\def\csname pct@rev.desc.d1.uiv.p1.Tsealed.grid2-d3.delta-tau-cross.tost5.p\endcsname{65.0}
\expandafter\def\csname ppt@rev.desc.d1.uiv.p1.Tsealed.grid2-d3.delta-tau-cross.tost5.p\endcsname{$+$65.0}
\expandafter\def\csname val@rev.desc.d1.uiv.p1.Tsealed.grid2-d3.material-flip\endcsname{0}
\expandafter\def\csname cnt@rev.desc.d1.uiv.p1.Tsealed.grid2-d3.material-flip\endcsname{0}
\expandafter\def\csname pct@rev.desc.d1.uiv.p1.Tsealed.grid2-d3.material-flip\endcsname{0.0}
\expandafter\def\csname ppt@rev.desc.d1.uiv.p1.Tsealed.grid2-d3.material-flip\endcsname{0.0}
\expandafter\def\csname val@rev.desc.d1.uiv.p1.Tsealed.grid2-d3.n-tasks\endcsname{144}
\expandafter\def\csname cnt@rev.desc.d1.uiv.p1.Tsealed.grid2-d3.n-tasks\endcsname{144}
\expandafter\def\csname pct@rev.desc.d1.uiv.p1.Tsealed.grid2-d3.n-tasks\endcsname{14400.0}
\expandafter\def\csname ppt@rev.desc.d1.uiv.p1.Tsealed.grid2-d3.n-tasks\endcsname{$+$14400.0}
\expandafter\def\csname val@rev.desc.d1.uiv.p1.Tsealed.grid2-d3.n-tasks-source-C\endcsname{96}
\expandafter\def\csname cnt@rev.desc.d1.uiv.p1.Tsealed.grid2-d3.n-tasks-source-C\endcsname{96}
\expandafter\def\csname pct@rev.desc.d1.uiv.p1.Tsealed.grid2-d3.n-tasks-source-C\endcsname{9600.0}
\expandafter\def\csname ppt@rev.desc.d1.uiv.p1.Tsealed.grid2-d3.n-tasks-source-C\endcsname{$+$9600.0}
\expandafter\def\csname val@rev.desc.d1.uiv.p1.Tsealed.grid2-d3.paper-tau-T\endcsname{-0.076}
\expandafter\def\csname cnt@rev.desc.d1.uiv.p1.Tsealed.grid2-d3.paper-tau-T\endcsname{0}
\expandafter\def\csname pct@rev.desc.d1.uiv.p1.Tsealed.grid2-d3.paper-tau-T\endcsname{$-$7.6}
\expandafter\def\csname ppt@rev.desc.d1.uiv.p1.Tsealed.grid2-d3.paper-tau-T\endcsname{$-$7.6}
\expandafter\def\csname val@rev.desc.d1.uiv.p1.Tsealed.grid2-d3.reproduces\endcsname{1}
\expandafter\def\csname cnt@rev.desc.d1.uiv.p1.Tsealed.grid2-d3.reproduces\endcsname{1}
\expandafter\def\csname pct@rev.desc.d1.uiv.p1.Tsealed.grid2-d3.reproduces\endcsname{100.0}
\expandafter\def\csname ppt@rev.desc.d1.uiv.p1.Tsealed.grid2-d3.reproduces\endcsname{$+$100.0}
\expandafter\def\csname val@rev.desc.d1.uiv.p1.Tsealed.grid2-d3.reproduces-ci\endcsname{1}
\expandafter\def\csname cnt@rev.desc.d1.uiv.p1.Tsealed.grid2-d3.reproduces-ci\endcsname{1}
\expandafter\def\csname pct@rev.desc.d1.uiv.p1.Tsealed.grid2-d3.reproduces-ci\endcsname{100.0}
\expandafter\def\csname ppt@rev.desc.d1.uiv.p1.Tsealed.grid2-d3.reproduces-ci\endcsname{$+$100.0}
\expandafter\def\csname val@rev.desc.d1.uiv.p1.Tsealed.grid2-d3.sign-differs\endcsname{0}
\expandafter\def\csname cnt@rev.desc.d1.uiv.p1.Tsealed.grid2-d3.sign-differs\endcsname{0}
\expandafter\def\csname pct@rev.desc.d1.uiv.p1.Tsealed.grid2-d3.sign-differs\endcsname{0.0}
\expandafter\def\csname ppt@rev.desc.d1.uiv.p1.Tsealed.grid2-d3.sign-differs\endcsname{0.0}
\expandafter\def\csname val@rev.desc.d1.uiv.p1.Tsealed.grid2-d3.tau-source-C.ci90.0\endcsname{-0.208}
\expandafter\def\csname cnt@rev.desc.d1.uiv.p1.Tsealed.grid2-d3.tau-source-C.ci90.0\endcsname{0}
\expandafter\def\csname pct@rev.desc.d1.uiv.p1.Tsealed.grid2-d3.tau-source-C.ci90.0\endcsname{$-$20.8}
\expandafter\def\csname ppt@rev.desc.d1.uiv.p1.Tsealed.grid2-d3.tau-source-C.ci90.0\endcsname{$-$20.8}
\expandafter\def\csname val@rev.desc.d1.uiv.p1.Tsealed.grid2-d3.tau-source-C.ci90.1\endcsname{-0.073}
\expandafter\def\csname cnt@rev.desc.d1.uiv.p1.Tsealed.grid2-d3.tau-source-C.ci90.1\endcsname{0}
\expandafter\def\csname pct@rev.desc.d1.uiv.p1.Tsealed.grid2-d3.tau-source-C.ci90.1\endcsname{$-$7.3}
\expandafter\def\csname ppt@rev.desc.d1.uiv.p1.Tsealed.grid2-d3.tau-source-C.ci90.1\endcsname{$-$7.3}
\expandafter\def\csname val@rev.desc.d1.uiv.p1.Tsealed.grid2-d3.tau-source-C.ci95-excludes-0\endcsname{1}
\expandafter\def\csname cnt@rev.desc.d1.uiv.p1.Tsealed.grid2-d3.tau-source-C.ci95-excludes-0\endcsname{1}
\expandafter\def\csname pct@rev.desc.d1.uiv.p1.Tsealed.grid2-d3.tau-source-C.ci95-excludes-0\endcsname{100.0}
\expandafter\def\csname ppt@rev.desc.d1.uiv.p1.Tsealed.grid2-d3.tau-source-C.ci95-excludes-0\endcsname{$+$100.0}
\expandafter\def\csname val@rev.desc.d1.uiv.p1.Tsealed.grid2-d3.tau-source-C.ci95.0\endcsname{-0.229}
\expandafter\def\csname cnt@rev.desc.d1.uiv.p1.Tsealed.grid2-d3.tau-source-C.ci95.0\endcsname{0}
\expandafter\def\csname pct@rev.desc.d1.uiv.p1.Tsealed.grid2-d3.tau-source-C.ci95.0\endcsname{$-$22.9}
\expandafter\def\csname ppt@rev.desc.d1.uiv.p1.Tsealed.grid2-d3.tau-source-C.ci95.0\endcsname{$-$22.9}
\expandafter\def\csname val@rev.desc.d1.uiv.p1.Tsealed.grid2-d3.tau-source-C.ci95.1\endcsname{-0.062}
\expandafter\def\csname cnt@rev.desc.d1.uiv.p1.Tsealed.grid2-d3.tau-source-C.ci95.1\endcsname{0}
\expandafter\def\csname pct@rev.desc.d1.uiv.p1.Tsealed.grid2-d3.tau-source-C.ci95.1\endcsname{$-$6.2}
\expandafter\def\csname ppt@rev.desc.d1.uiv.p1.Tsealed.grid2-d3.tau-source-C.ci95.1\endcsname{$-$6.2}
\expandafter\def\csname val@rev.desc.d1.uiv.p1.Tsealed.grid2-d3.tau-source-C.mean\endcsname{-0.146}
\expandafter\def\csname cnt@rev.desc.d1.uiv.p1.Tsealed.grid2-d3.tau-source-C.mean\endcsname{0}
\expandafter\def\csname pct@rev.desc.d1.uiv.p1.Tsealed.grid2-d3.tau-source-C.mean\endcsname{$-$14.6}
\expandafter\def\csname ppt@rev.desc.d1.uiv.p1.Tsealed.grid2-d3.tau-source-C.mean\endcsname{$-$14.6}
\expandafter\def\csname val@rev.desc.d1.uiv.p1.Tsealed.grid2-d3.tau-source-C.n-apps\endcsname{12}
\expandafter\def\csname cnt@rev.desc.d1.uiv.p1.Tsealed.grid2-d3.tau-source-C.n-apps\endcsname{12}
\expandafter\def\csname pct@rev.desc.d1.uiv.p1.Tsealed.grid2-d3.tau-source-C.n-apps\endcsname{1200.0}
\expandafter\def\csname ppt@rev.desc.d1.uiv.p1.Tsealed.grid2-d3.tau-source-C.n-apps\endcsname{$+$1200.0}
\expandafter\def\csname val@rev.desc.d1.uiv.p1.Tsealed.grid2-d3.tau-source-C.tost1.ci90-inside\endcsname{0}
\expandafter\def\csname cnt@rev.desc.d1.uiv.p1.Tsealed.grid2-d3.tau-source-C.tost1.ci90-inside\endcsname{0}
\expandafter\def\csname pct@rev.desc.d1.uiv.p1.Tsealed.grid2-d3.tau-source-C.tost1.ci90-inside\endcsname{0.0}
\expandafter\def\csname ppt@rev.desc.d1.uiv.p1.Tsealed.grid2-d3.tau-source-C.tost1.ci90-inside\endcsname{0.0}
\expandafter\def\csname val@rev.desc.d1.uiv.p1.Tsealed.grid2-d3.tau-source-C.tost1.equivalent\endcsname{0}
\expandafter\def\csname cnt@rev.desc.d1.uiv.p1.Tsealed.grid2-d3.tau-source-C.tost1.equivalent\endcsname{0}
\expandafter\def\csname pct@rev.desc.d1.uiv.p1.Tsealed.grid2-d3.tau-source-C.tost1.equivalent\endcsname{0.0}
\expandafter\def\csname ppt@rev.desc.d1.uiv.p1.Tsealed.grid2-d3.tau-source-C.tost1.equivalent\endcsname{0.0}
\expandafter\def\csname val@rev.desc.d1.uiv.p1.Tsealed.grid2-d3.tau-source-C.tost1.p\endcsname{0.999}
\expandafter\def\csname cnt@rev.desc.d1.uiv.p1.Tsealed.grid2-d3.tau-source-C.tost1.p\endcsname{1}
\expandafter\def\csname pct@rev.desc.d1.uiv.p1.Tsealed.grid2-d3.tau-source-C.tost1.p\endcsname{99.9}
\expandafter\def\csname ppt@rev.desc.d1.uiv.p1.Tsealed.grid2-d3.tau-source-C.tost1.p\endcsname{$+$99.9}
\expandafter\def\csname val@rev.desc.d1.uiv.p1.Tsealed.grid2-d3.tau-source-C.tost5.ci90-inside\endcsname{0}
\expandafter\def\csname cnt@rev.desc.d1.uiv.p1.Tsealed.grid2-d3.tau-source-C.tost5.ci90-inside\endcsname{0}
\expandafter\def\csname pct@rev.desc.d1.uiv.p1.Tsealed.grid2-d3.tau-source-C.tost5.ci90-inside\endcsname{0.0}
\expandafter\def\csname ppt@rev.desc.d1.uiv.p1.Tsealed.grid2-d3.tau-source-C.tost5.ci90-inside\endcsname{0.0}
\expandafter\def\csname val@rev.desc.d1.uiv.p1.Tsealed.grid2-d3.tau-source-C.tost5.equivalent\endcsname{0}
\expandafter\def\csname cnt@rev.desc.d1.uiv.p1.Tsealed.grid2-d3.tau-source-C.tost5.equivalent\endcsname{0}
\expandafter\def\csname pct@rev.desc.d1.uiv.p1.Tsealed.grid2-d3.tau-source-C.tost5.equivalent\endcsname{0.0}
\expandafter\def\csname ppt@rev.desc.d1.uiv.p1.Tsealed.grid2-d3.tau-source-C.tost5.equivalent\endcsname{0.0}
\expandafter\def\csname val@rev.desc.d1.uiv.p1.Tsealed.grid2-d3.tau-source-C.tost5.p\endcsname{0.992}
\expandafter\def\csname cnt@rev.desc.d1.uiv.p1.Tsealed.grid2-d3.tau-source-C.tost5.p\endcsname{1}
\expandafter\def\csname pct@rev.desc.d1.uiv.p1.Tsealed.grid2-d3.tau-source-C.tost5.p\endcsname{99.2}
\expandafter\def\csname ppt@rev.desc.d1.uiv.p1.Tsealed.grid2-d3.tau-source-C.tost5.p\endcsname{$+$99.2}
\expandafter\def\csname val@rev.desc.d1.uiv.p1.Tsealed.grid2-d3.tau-target-T.ci90.0\endcsname{-0.125}
\expandafter\def\csname cnt@rev.desc.d1.uiv.p1.Tsealed.grid2-d3.tau-target-T.ci90.0\endcsname{0}
\expandafter\def\csname pct@rev.desc.d1.uiv.p1.Tsealed.grid2-d3.tau-target-T.ci90.0\endcsname{$-$12.5}
\expandafter\def\csname ppt@rev.desc.d1.uiv.p1.Tsealed.grid2-d3.tau-target-T.ci90.0\endcsname{$-$12.5}
\expandafter\def\csname val@rev.desc.d1.uiv.p1.Tsealed.grid2-d3.tau-target-T.ci90.1\endcsname{-0.028}
\expandafter\def\csname cnt@rev.desc.d1.uiv.p1.Tsealed.grid2-d3.tau-target-T.ci90.1\endcsname{0}
\expandafter\def\csname pct@rev.desc.d1.uiv.p1.Tsealed.grid2-d3.tau-target-T.ci90.1\endcsname{$-$2.8}
\expandafter\def\csname ppt@rev.desc.d1.uiv.p1.Tsealed.grid2-d3.tau-target-T.ci90.1\endcsname{$-$2.8}
\expandafter\def\csname val@rev.desc.d1.uiv.p1.Tsealed.grid2-d3.tau-target-T.ci95-excludes-0\endcsname{1}
\expandafter\def\csname cnt@rev.desc.d1.uiv.p1.Tsealed.grid2-d3.tau-target-T.ci95-excludes-0\endcsname{1}
\expandafter\def\csname pct@rev.desc.d1.uiv.p1.Tsealed.grid2-d3.tau-target-T.ci95-excludes-0\endcsname{100.0}
\expandafter\def\csname ppt@rev.desc.d1.uiv.p1.Tsealed.grid2-d3.tau-target-T.ci95-excludes-0\endcsname{$+$100.0}
\expandafter\def\csname val@rev.desc.d1.uiv.p1.Tsealed.grid2-d3.tau-target-T.ci95.0\endcsname{-0.139}
\expandafter\def\csname cnt@rev.desc.d1.uiv.p1.Tsealed.grid2-d3.tau-target-T.ci95.0\endcsname{0}
\expandafter\def\csname pct@rev.desc.d1.uiv.p1.Tsealed.grid2-d3.tau-target-T.ci95.0\endcsname{$-$13.9}
\expandafter\def\csname ppt@rev.desc.d1.uiv.p1.Tsealed.grid2-d3.tau-target-T.ci95.0\endcsname{$-$13.9}
\expandafter\def\csname val@rev.desc.d1.uiv.p1.Tsealed.grid2-d3.tau-target-T.ci95.1\endcsname{-0.021}
\expandafter\def\csname cnt@rev.desc.d1.uiv.p1.Tsealed.grid2-d3.tau-target-T.ci95.1\endcsname{0}
\expandafter\def\csname pct@rev.desc.d1.uiv.p1.Tsealed.grid2-d3.tau-target-T.ci95.1\endcsname{$-$2.1}
\expandafter\def\csname ppt@rev.desc.d1.uiv.p1.Tsealed.grid2-d3.tau-target-T.ci95.1\endcsname{$-$2.1}
\expandafter\def\csname val@rev.desc.d1.uiv.p1.Tsealed.grid2-d3.tau-target-T.mean\endcsname{-0.076}
\expandafter\def\csname cnt@rev.desc.d1.uiv.p1.Tsealed.grid2-d3.tau-target-T.mean\endcsname{0}
\expandafter\def\csname pct@rev.desc.d1.uiv.p1.Tsealed.grid2-d3.tau-target-T.mean\endcsname{$-$7.6}
\expandafter\def\csname ppt@rev.desc.d1.uiv.p1.Tsealed.grid2-d3.tau-target-T.mean\endcsname{$-$7.6}
\expandafter\def\csname val@rev.desc.d1.uiv.p1.Tsealed.grid2-d3.tau-target-T.n-apps\endcsname{24}
\expandafter\def\csname cnt@rev.desc.d1.uiv.p1.Tsealed.grid2-d3.tau-target-T.n-apps\endcsname{24}
\expandafter\def\csname pct@rev.desc.d1.uiv.p1.Tsealed.grid2-d3.tau-target-T.n-apps\endcsname{2400.0}
\expandafter\def\csname ppt@rev.desc.d1.uiv.p1.Tsealed.grid2-d3.tau-target-T.n-apps\endcsname{$+$2400.0}
\expandafter\def\csname val@rev.desc.d1.uiv.p1.Tsealed.grid2-d3.tau-target-T.tost1.ci90-inside\endcsname{0}
\expandafter\def\csname cnt@rev.desc.d1.uiv.p1.Tsealed.grid2-d3.tau-target-T.tost1.ci90-inside\endcsname{0}
\expandafter\def\csname pct@rev.desc.d1.uiv.p1.Tsealed.grid2-d3.tau-target-T.tost1.ci90-inside\endcsname{0.0}
\expandafter\def\csname ppt@rev.desc.d1.uiv.p1.Tsealed.grid2-d3.tau-target-T.tost1.ci90-inside\endcsname{0.0}
\expandafter\def\csname val@rev.desc.d1.uiv.p1.Tsealed.grid2-d3.tau-target-T.tost1.equivalent\endcsname{0}
\expandafter\def\csname cnt@rev.desc.d1.uiv.p1.Tsealed.grid2-d3.tau-target-T.tost1.equivalent\endcsname{0}
\expandafter\def\csname pct@rev.desc.d1.uiv.p1.Tsealed.grid2-d3.tau-target-T.tost1.equivalent\endcsname{0.0}
\expandafter\def\csname ppt@rev.desc.d1.uiv.p1.Tsealed.grid2-d3.tau-target-T.tost1.equivalent\endcsname{0.0}
\expandafter\def\csname val@rev.desc.d1.uiv.p1.Tsealed.grid2-d3.tau-target-T.tost1.p\endcsname{0.986}
\expandafter\def\csname cnt@rev.desc.d1.uiv.p1.Tsealed.grid2-d3.tau-target-T.tost1.p\endcsname{1}
\expandafter\def\csname pct@rev.desc.d1.uiv.p1.Tsealed.grid2-d3.tau-target-T.tost1.p\endcsname{98.6}
\expandafter\def\csname ppt@rev.desc.d1.uiv.p1.Tsealed.grid2-d3.tau-target-T.tost1.p\endcsname{$+$98.6}
\expandafter\def\csname val@rev.desc.d1.uiv.p1.Tsealed.grid2-d3.tau-target-T.tost5.ci90-inside\endcsname{0}
\expandafter\def\csname cnt@rev.desc.d1.uiv.p1.Tsealed.grid2-d3.tau-target-T.tost5.ci90-inside\endcsname{0}
\expandafter\def\csname pct@rev.desc.d1.uiv.p1.Tsealed.grid2-d3.tau-target-T.tost5.ci90-inside\endcsname{0.0}
\expandafter\def\csname ppt@rev.desc.d1.uiv.p1.Tsealed.grid2-d3.tau-target-T.tost5.ci90-inside\endcsname{0.0}
\expandafter\def\csname val@rev.desc.d1.uiv.p1.Tsealed.grid2-d3.tau-target-T.tost5.equivalent\endcsname{0}
\expandafter\def\csname cnt@rev.desc.d1.uiv.p1.Tsealed.grid2-d3.tau-target-T.tost5.equivalent\endcsname{0}
\expandafter\def\csname pct@rev.desc.d1.uiv.p1.Tsealed.grid2-d3.tau-target-T.tost5.equivalent\endcsname{0.0}
\expandafter\def\csname ppt@rev.desc.d1.uiv.p1.Tsealed.grid2-d3.tau-target-T.tost5.equivalent\endcsname{0.0}
\expandafter\def\csname val@rev.desc.d1.uiv.p1.Tsealed.grid2-d3.tau-target-T.tost5.p\endcsname{0.788}
\expandafter\def\csname cnt@rev.desc.d1.uiv.p1.Tsealed.grid2-d3.tau-target-T.tost5.p\endcsname{1}
\expandafter\def\csname pct@rev.desc.d1.uiv.p1.Tsealed.grid2-d3.tau-target-T.tost5.p\endcsname{78.8}
\expandafter\def\csname ppt@rev.desc.d1.uiv.p1.Tsealed.grid2-d3.tau-target-T.tost5.p\endcsname{$+$78.8}
\expandafter\def\csname val@rev.desc.d1.uiv.p1.Tsealed.grid2-d3.zero-estimate\endcsname{0}
\expandafter\def\csname cnt@rev.desc.d1.uiv.p1.Tsealed.grid2-d3.zero-estimate\endcsname{0}
\expandafter\def\csname pct@rev.desc.d1.uiv.p1.Tsealed.grid2-d3.zero-estimate\endcsname{0.0}
\expandafter\def\csname ppt@rev.desc.d1.uiv.p1.Tsealed.grid2-d3.zero-estimate\endcsname{0.0}
\expandafter\def\csname val@rev.desc.d1.uiv.p1.Tsealed.zoomclick-d3.delta-tau-cross.ci90.0\endcsname{-0.087}
\expandafter\def\csname cnt@rev.desc.d1.uiv.p1.Tsealed.zoomclick-d3.delta-tau-cross.ci90.0\endcsname{0}
\expandafter\def\csname pct@rev.desc.d1.uiv.p1.Tsealed.zoomclick-d3.delta-tau-cross.ci90.0\endcsname{$-$8.7}
\expandafter\def\csname ppt@rev.desc.d1.uiv.p1.Tsealed.zoomclick-d3.delta-tau-cross.ci90.0\endcsname{$-$8.7}
\expandafter\def\csname val@rev.desc.d1.uiv.p1.Tsealed.zoomclick-d3.delta-tau-cross.ci90.1\endcsname{0.038}
\expandafter\def\csname cnt@rev.desc.d1.uiv.p1.Tsealed.zoomclick-d3.delta-tau-cross.ci90.1\endcsname{0}
\expandafter\def\csname pct@rev.desc.d1.uiv.p1.Tsealed.zoomclick-d3.delta-tau-cross.ci90.1\endcsname{3.8}
\expandafter\def\csname ppt@rev.desc.d1.uiv.p1.Tsealed.zoomclick-d3.delta-tau-cross.ci90.1\endcsname{$+$3.8}
\expandafter\def\csname val@rev.desc.d1.uiv.p1.Tsealed.zoomclick-d3.delta-tau-cross.ci95-excludes-0\endcsname{0}
\expandafter\def\csname cnt@rev.desc.d1.uiv.p1.Tsealed.zoomclick-d3.delta-tau-cross.ci95-excludes-0\endcsname{0}
\expandafter\def\csname pct@rev.desc.d1.uiv.p1.Tsealed.zoomclick-d3.delta-tau-cross.ci95-excludes-0\endcsname{0.0}
\expandafter\def\csname ppt@rev.desc.d1.uiv.p1.Tsealed.zoomclick-d3.delta-tau-cross.ci95-excludes-0\endcsname{0.0}
\expandafter\def\csname val@rev.desc.d1.uiv.p1.Tsealed.zoomclick-d3.delta-tau-cross.ci95.0\endcsname{-0.101}
\expandafter\def\csname cnt@rev.desc.d1.uiv.p1.Tsealed.zoomclick-d3.delta-tau-cross.ci95.0\endcsname{0}
\expandafter\def\csname pct@rev.desc.d1.uiv.p1.Tsealed.zoomclick-d3.delta-tau-cross.ci95.0\endcsname{$-$10.1}
\expandafter\def\csname ppt@rev.desc.d1.uiv.p1.Tsealed.zoomclick-d3.delta-tau-cross.ci95.0\endcsname{$-$10.1}
\expandafter\def\csname val@rev.desc.d1.uiv.p1.Tsealed.zoomclick-d3.delta-tau-cross.ci95.1\endcsname{0.049}
\expandafter\def\csname cnt@rev.desc.d1.uiv.p1.Tsealed.zoomclick-d3.delta-tau-cross.ci95.1\endcsname{0}
\expandafter\def\csname pct@rev.desc.d1.uiv.p1.Tsealed.zoomclick-d3.delta-tau-cross.ci95.1\endcsname{4.9}
\expandafter\def\csname ppt@rev.desc.d1.uiv.p1.Tsealed.zoomclick-d3.delta-tau-cross.ci95.1\endcsname{$+$4.9}
\expandafter\def\csname val@rev.desc.d1.uiv.p1.Tsealed.zoomclick-d3.delta-tau-cross.mean\endcsname{-0.024}
\expandafter\def\csname cnt@rev.desc.d1.uiv.p1.Tsealed.zoomclick-d3.delta-tau-cross.mean\endcsname{0}
\expandafter\def\csname pct@rev.desc.d1.uiv.p1.Tsealed.zoomclick-d3.delta-tau-cross.mean\endcsname{$-$2.4}
\expandafter\def\csname ppt@rev.desc.d1.uiv.p1.Tsealed.zoomclick-d3.delta-tau-cross.mean\endcsname{$-$2.4}
\expandafter\def\csname val@rev.desc.d1.uiv.p1.Tsealed.zoomclick-d3.delta-tau-cross.n-apps.0\endcsname{24}
\expandafter\def\csname cnt@rev.desc.d1.uiv.p1.Tsealed.zoomclick-d3.delta-tau-cross.n-apps.0\endcsname{24}
\expandafter\def\csname pct@rev.desc.d1.uiv.p1.Tsealed.zoomclick-d3.delta-tau-cross.n-apps.0\endcsname{2400.0}
\expandafter\def\csname ppt@rev.desc.d1.uiv.p1.Tsealed.zoomclick-d3.delta-tau-cross.n-apps.0\endcsname{$+$2400.0}
\expandafter\def\csname val@rev.desc.d1.uiv.p1.Tsealed.zoomclick-d3.delta-tau-cross.n-apps.1\endcsname{12}
\expandafter\def\csname cnt@rev.desc.d1.uiv.p1.Tsealed.zoomclick-d3.delta-tau-cross.n-apps.1\endcsname{12}
\expandafter\def\csname pct@rev.desc.d1.uiv.p1.Tsealed.zoomclick-d3.delta-tau-cross.n-apps.1\endcsname{1200.0}
\expandafter\def\csname ppt@rev.desc.d1.uiv.p1.Tsealed.zoomclick-d3.delta-tau-cross.n-apps.1\endcsname{$+$1200.0}
\expandafter\def\csname val@rev.desc.d1.uiv.p1.Tsealed.zoomclick-d3.delta-tau-cross.tost1.ci90-inside\endcsname{0}
\expandafter\def\csname cnt@rev.desc.d1.uiv.p1.Tsealed.zoomclick-d3.delta-tau-cross.tost1.ci90-inside\endcsname{0}
\expandafter\def\csname pct@rev.desc.d1.uiv.p1.Tsealed.zoomclick-d3.delta-tau-cross.tost1.ci90-inside\endcsname{0.0}
\expandafter\def\csname ppt@rev.desc.d1.uiv.p1.Tsealed.zoomclick-d3.delta-tau-cross.tost1.ci90-inside\endcsname{0.0}
\expandafter\def\csname val@rev.desc.d1.uiv.p1.Tsealed.zoomclick-d3.delta-tau-cross.tost1.equivalent\endcsname{0}
\expandafter\def\csname cnt@rev.desc.d1.uiv.p1.Tsealed.zoomclick-d3.delta-tau-cross.tost1.equivalent\endcsname{0}
\expandafter\def\csname pct@rev.desc.d1.uiv.p1.Tsealed.zoomclick-d3.delta-tau-cross.tost1.equivalent\endcsname{0.0}
\expandafter\def\csname ppt@rev.desc.d1.uiv.p1.Tsealed.zoomclick-d3.delta-tau-cross.tost1.equivalent\endcsname{0.0}
\expandafter\def\csname val@rev.desc.d1.uiv.p1.Tsealed.zoomclick-d3.delta-tau-cross.tost1.p\endcsname{0.662}
\expandafter\def\csname cnt@rev.desc.d1.uiv.p1.Tsealed.zoomclick-d3.delta-tau-cross.tost1.p\endcsname{1}
\expandafter\def\csname pct@rev.desc.d1.uiv.p1.Tsealed.zoomclick-d3.delta-tau-cross.tost1.p\endcsname{66.2}
\expandafter\def\csname ppt@rev.desc.d1.uiv.p1.Tsealed.zoomclick-d3.delta-tau-cross.tost1.p\endcsname{$+$66.2}
\expandafter\def\csname val@rev.desc.d1.uiv.p1.Tsealed.zoomclick-d3.delta-tau-cross.tost5.ci90-inside\endcsname{0}
\expandafter\def\csname cnt@rev.desc.d1.uiv.p1.Tsealed.zoomclick-d3.delta-tau-cross.tost5.ci90-inside\endcsname{0}
\expandafter\def\csname pct@rev.desc.d1.uiv.p1.Tsealed.zoomclick-d3.delta-tau-cross.tost5.ci90-inside\endcsname{0.0}
\expandafter\def\csname ppt@rev.desc.d1.uiv.p1.Tsealed.zoomclick-d3.delta-tau-cross.tost5.ci90-inside\endcsname{0.0}
\expandafter\def\csname val@rev.desc.d1.uiv.p1.Tsealed.zoomclick-d3.delta-tau-cross.tost5.equivalent\endcsname{0}
\expandafter\def\csname cnt@rev.desc.d1.uiv.p1.Tsealed.zoomclick-d3.delta-tau-cross.tost5.equivalent\endcsname{0}
\expandafter\def\csname pct@rev.desc.d1.uiv.p1.Tsealed.zoomclick-d3.delta-tau-cross.tost5.equivalent\endcsname{0.0}
\expandafter\def\csname ppt@rev.desc.d1.uiv.p1.Tsealed.zoomclick-d3.delta-tau-cross.tost5.equivalent\endcsname{0.0}
\expandafter\def\csname val@rev.desc.d1.uiv.p1.Tsealed.zoomclick-d3.delta-tau-cross.tost5.p\endcsname{0.235}
\expandafter\def\csname cnt@rev.desc.d1.uiv.p1.Tsealed.zoomclick-d3.delta-tau-cross.tost5.p\endcsname{0}
\expandafter\def\csname pct@rev.desc.d1.uiv.p1.Tsealed.zoomclick-d3.delta-tau-cross.tost5.p\endcsname{23.5}
\expandafter\def\csname ppt@rev.desc.d1.uiv.p1.Tsealed.zoomclick-d3.delta-tau-cross.tost5.p\endcsname{$+$23.5}
\expandafter\def\csname val@rev.desc.d1.uiv.p1.Tsealed.zoomclick-d3.material-flip\endcsname{0}
\expandafter\def\csname cnt@rev.desc.d1.uiv.p1.Tsealed.zoomclick-d3.material-flip\endcsname{0}
\expandafter\def\csname pct@rev.desc.d1.uiv.p1.Tsealed.zoomclick-d3.material-flip\endcsname{0.0}
\expandafter\def\csname ppt@rev.desc.d1.uiv.p1.Tsealed.zoomclick-d3.material-flip\endcsname{0.0}
\expandafter\def\csname val@rev.desc.d1.uiv.p1.Tsealed.zoomclick-d3.n-tasks\endcsname{144}
\expandafter\def\csname cnt@rev.desc.d1.uiv.p1.Tsealed.zoomclick-d3.n-tasks\endcsname{144}
\expandafter\def\csname pct@rev.desc.d1.uiv.p1.Tsealed.zoomclick-d3.n-tasks\endcsname{14400.0}
\expandafter\def\csname ppt@rev.desc.d1.uiv.p1.Tsealed.zoomclick-d3.n-tasks\endcsname{$+$14400.0}
\expandafter\def\csname val@rev.desc.d1.uiv.p1.Tsealed.zoomclick-d3.n-tasks-source-C\endcsname{96}
\expandafter\def\csname cnt@rev.desc.d1.uiv.p1.Tsealed.zoomclick-d3.n-tasks-source-C\endcsname{96}
\expandafter\def\csname pct@rev.desc.d1.uiv.p1.Tsealed.zoomclick-d3.n-tasks-source-C\endcsname{9600.0}
\expandafter\def\csname ppt@rev.desc.d1.uiv.p1.Tsealed.zoomclick-d3.n-tasks-source-C\endcsname{$+$9600.0}
\expandafter\def\csname val@rev.desc.d1.uiv.p1.Tsealed.zoomclick-d3.paper-tau-T\endcsname{-0.056}
\expandafter\def\csname cnt@rev.desc.d1.uiv.p1.Tsealed.zoomclick-d3.paper-tau-T\endcsname{0}
\expandafter\def\csname pct@rev.desc.d1.uiv.p1.Tsealed.zoomclick-d3.paper-tau-T\endcsname{$-$5.6}
\expandafter\def\csname ppt@rev.desc.d1.uiv.p1.Tsealed.zoomclick-d3.paper-tau-T\endcsname{$-$5.6}
\expandafter\def\csname val@rev.desc.d1.uiv.p1.Tsealed.zoomclick-d3.reproduces\endcsname{1}
\expandafter\def\csname cnt@rev.desc.d1.uiv.p1.Tsealed.zoomclick-d3.reproduces\endcsname{1}
\expandafter\def\csname pct@rev.desc.d1.uiv.p1.Tsealed.zoomclick-d3.reproduces\endcsname{100.0}
\expandafter\def\csname ppt@rev.desc.d1.uiv.p1.Tsealed.zoomclick-d3.reproduces\endcsname{$+$100.0}
\expandafter\def\csname val@rev.desc.d1.uiv.p1.Tsealed.zoomclick-d3.reproduces-ci\endcsname{1}
\expandafter\def\csname cnt@rev.desc.d1.uiv.p1.Tsealed.zoomclick-d3.reproduces-ci\endcsname{1}
\expandafter\def\csname pct@rev.desc.d1.uiv.p1.Tsealed.zoomclick-d3.reproduces-ci\endcsname{100.0}
\expandafter\def\csname ppt@rev.desc.d1.uiv.p1.Tsealed.zoomclick-d3.reproduces-ci\endcsname{$+$100.0}
\expandafter\def\csname val@rev.desc.d1.uiv.p1.Tsealed.zoomclick-d3.sign-differs\endcsname{0}
\expandafter\def\csname cnt@rev.desc.d1.uiv.p1.Tsealed.zoomclick-d3.sign-differs\endcsname{0}
\expandafter\def\csname pct@rev.desc.d1.uiv.p1.Tsealed.zoomclick-d3.sign-differs\endcsname{0.0}
\expandafter\def\csname ppt@rev.desc.d1.uiv.p1.Tsealed.zoomclick-d3.sign-differs\endcsname{0.0}
\expandafter\def\csname val@rev.desc.d1.uiv.p1.Tsealed.zoomclick-d3.tau-source-C.ci90.0\endcsname{-0.073}
\expandafter\def\csname cnt@rev.desc.d1.uiv.p1.Tsealed.zoomclick-d3.tau-source-C.ci90.0\endcsname{0}
\expandafter\def\csname pct@rev.desc.d1.uiv.p1.Tsealed.zoomclick-d3.tau-source-C.ci90.0\endcsname{$-$7.3}
\expandafter\def\csname ppt@rev.desc.d1.uiv.p1.Tsealed.zoomclick-d3.tau-source-C.ci90.0\endcsname{$-$7.3}
\expandafter\def\csname val@rev.desc.d1.uiv.p1.Tsealed.zoomclick-d3.tau-source-C.ci90.1\endcsname{0.010}
\expandafter\def\csname cnt@rev.desc.d1.uiv.p1.Tsealed.zoomclick-d3.tau-source-C.ci90.1\endcsname{0}
\expandafter\def\csname pct@rev.desc.d1.uiv.p1.Tsealed.zoomclick-d3.tau-source-C.ci90.1\endcsname{1.0}
\expandafter\def\csname ppt@rev.desc.d1.uiv.p1.Tsealed.zoomclick-d3.tau-source-C.ci90.1\endcsname{$+$1.0}
\expandafter\def\csname val@rev.desc.d1.uiv.p1.Tsealed.zoomclick-d3.tau-source-C.ci95-excludes-0\endcsname{0}
\expandafter\def\csname cnt@rev.desc.d1.uiv.p1.Tsealed.zoomclick-d3.tau-source-C.ci95-excludes-0\endcsname{0}
\expandafter\def\csname pct@rev.desc.d1.uiv.p1.Tsealed.zoomclick-d3.tau-source-C.ci95-excludes-0\endcsname{0.0}
\expandafter\def\csname ppt@rev.desc.d1.uiv.p1.Tsealed.zoomclick-d3.tau-source-C.ci95-excludes-0\endcsname{0.0}
\expandafter\def\csname val@rev.desc.d1.uiv.p1.Tsealed.zoomclick-d3.tau-source-C.ci95.0\endcsname{-0.083}
\expandafter\def\csname cnt@rev.desc.d1.uiv.p1.Tsealed.zoomclick-d3.tau-source-C.ci95.0\endcsname{0}
\expandafter\def\csname pct@rev.desc.d1.uiv.p1.Tsealed.zoomclick-d3.tau-source-C.ci95.0\endcsname{$-$8.3}
\expandafter\def\csname ppt@rev.desc.d1.uiv.p1.Tsealed.zoomclick-d3.tau-source-C.ci95.0\endcsname{$-$8.3}
\expandafter\def\csname val@rev.desc.d1.uiv.p1.Tsealed.zoomclick-d3.tau-source-C.ci95.1\endcsname{0.021}
\expandafter\def\csname cnt@rev.desc.d1.uiv.p1.Tsealed.zoomclick-d3.tau-source-C.ci95.1\endcsname{0}
\expandafter\def\csname pct@rev.desc.d1.uiv.p1.Tsealed.zoomclick-d3.tau-source-C.ci95.1\endcsname{2.1}
\expandafter\def\csname ppt@rev.desc.d1.uiv.p1.Tsealed.zoomclick-d3.tau-source-C.ci95.1\endcsname{$+$2.1}
\expandafter\def\csname val@rev.desc.d1.uiv.p1.Tsealed.zoomclick-d3.tau-source-C.mean\endcsname{-0.031}
\expandafter\def\csname cnt@rev.desc.d1.uiv.p1.Tsealed.zoomclick-d3.tau-source-C.mean\endcsname{0}
\expandafter\def\csname pct@rev.desc.d1.uiv.p1.Tsealed.zoomclick-d3.tau-source-C.mean\endcsname{$-$3.1}
\expandafter\def\csname ppt@rev.desc.d1.uiv.p1.Tsealed.zoomclick-d3.tau-source-C.mean\endcsname{$-$3.1}
\expandafter\def\csname val@rev.desc.d1.uiv.p1.Tsealed.zoomclick-d3.tau-source-C.n-apps\endcsname{12}
\expandafter\def\csname cnt@rev.desc.d1.uiv.p1.Tsealed.zoomclick-d3.tau-source-C.n-apps\endcsname{12}
\expandafter\def\csname pct@rev.desc.d1.uiv.p1.Tsealed.zoomclick-d3.tau-source-C.n-apps\endcsname{1200.0}
\expandafter\def\csname ppt@rev.desc.d1.uiv.p1.Tsealed.zoomclick-d3.tau-source-C.n-apps\endcsname{$+$1200.0}
\expandafter\def\csname val@rev.desc.d1.uiv.p1.Tsealed.zoomclick-d3.tau-source-C.tost1.ci90-inside\endcsname{0}
\expandafter\def\csname cnt@rev.desc.d1.uiv.p1.Tsealed.zoomclick-d3.tau-source-C.tost1.ci90-inside\endcsname{0}
\expandafter\def\csname pct@rev.desc.d1.uiv.p1.Tsealed.zoomclick-d3.tau-source-C.tost1.ci90-inside\endcsname{0.0}
\expandafter\def\csname ppt@rev.desc.d1.uiv.p1.Tsealed.zoomclick-d3.tau-source-C.tost1.ci90-inside\endcsname{0.0}
\expandafter\def\csname val@rev.desc.d1.uiv.p1.Tsealed.zoomclick-d3.tau-source-C.tost1.equivalent\endcsname{0}
\expandafter\def\csname cnt@rev.desc.d1.uiv.p1.Tsealed.zoomclick-d3.tau-source-C.tost1.equivalent\endcsname{0}
\expandafter\def\csname pct@rev.desc.d1.uiv.p1.Tsealed.zoomclick-d3.tau-source-C.tost1.equivalent\endcsname{0.0}
\expandafter\def\csname ppt@rev.desc.d1.uiv.p1.Tsealed.zoomclick-d3.tau-source-C.tost1.equivalent\endcsname{0.0}
\expandafter\def\csname val@rev.desc.d1.uiv.p1.Tsealed.zoomclick-d3.tau-source-C.tost1.p\endcsname{0.840}
\expandafter\def\csname cnt@rev.desc.d1.uiv.p1.Tsealed.zoomclick-d3.tau-source-C.tost1.p\endcsname{1}
\expandafter\def\csname pct@rev.desc.d1.uiv.p1.Tsealed.zoomclick-d3.tau-source-C.tost1.p\endcsname{84.0}
\expandafter\def\csname ppt@rev.desc.d1.uiv.p1.Tsealed.zoomclick-d3.tau-source-C.tost1.p\endcsname{$+$84.0}
\expandafter\def\csname val@rev.desc.d1.uiv.p1.Tsealed.zoomclick-d3.tau-source-C.tost5.ci90-inside\endcsname{0}
\expandafter\def\csname cnt@rev.desc.d1.uiv.p1.Tsealed.zoomclick-d3.tau-source-C.tost5.ci90-inside\endcsname{0}
\expandafter\def\csname pct@rev.desc.d1.uiv.p1.Tsealed.zoomclick-d3.tau-source-C.tost5.ci90-inside\endcsname{0.0}
\expandafter\def\csname ppt@rev.desc.d1.uiv.p1.Tsealed.zoomclick-d3.tau-source-C.tost5.ci90-inside\endcsname{0.0}
\expandafter\def\csname val@rev.desc.d1.uiv.p1.Tsealed.zoomclick-d3.tau-source-C.tost5.equivalent\endcsname{0}
\expandafter\def\csname cnt@rev.desc.d1.uiv.p1.Tsealed.zoomclick-d3.tau-source-C.tost5.equivalent\endcsname{0}
\expandafter\def\csname pct@rev.desc.d1.uiv.p1.Tsealed.zoomclick-d3.tau-source-C.tost5.equivalent\endcsname{0.0}
\expandafter\def\csname ppt@rev.desc.d1.uiv.p1.Tsealed.zoomclick-d3.tau-source-C.tost5.equivalent\endcsname{0.0}
\expandafter\def\csname val@rev.desc.d1.uiv.p1.Tsealed.zoomclick-d3.tau-source-C.tost5.p\endcsname{0.264}
\expandafter\def\csname cnt@rev.desc.d1.uiv.p1.Tsealed.zoomclick-d3.tau-source-C.tost5.p\endcsname{0}
\expandafter\def\csname pct@rev.desc.d1.uiv.p1.Tsealed.zoomclick-d3.tau-source-C.tost5.p\endcsname{26.4}
\expandafter\def\csname ppt@rev.desc.d1.uiv.p1.Tsealed.zoomclick-d3.tau-source-C.tost5.p\endcsname{$+$26.4}
\expandafter\def\csname val@rev.desc.d1.uiv.p1.Tsealed.zoomclick-d3.tau-target-T.ci90.0\endcsname{-0.104}
\expandafter\def\csname cnt@rev.desc.d1.uiv.p1.Tsealed.zoomclick-d3.tau-target-T.ci90.0\endcsname{0}
\expandafter\def\csname pct@rev.desc.d1.uiv.p1.Tsealed.zoomclick-d3.tau-target-T.ci90.0\endcsname{$-$10.4}
\expandafter\def\csname ppt@rev.desc.d1.uiv.p1.Tsealed.zoomclick-d3.tau-target-T.ci90.0\endcsname{$-$10.4}
\expandafter\def\csname val@rev.desc.d1.uiv.p1.Tsealed.zoomclick-d3.tau-target-T.ci90.1\endcsname{-0.014}
\expandafter\def\csname cnt@rev.desc.d1.uiv.p1.Tsealed.zoomclick-d3.tau-target-T.ci90.1\endcsname{0}
\expandafter\def\csname pct@rev.desc.d1.uiv.p1.Tsealed.zoomclick-d3.tau-target-T.ci90.1\endcsname{$-$1.4}
\expandafter\def\csname ppt@rev.desc.d1.uiv.p1.Tsealed.zoomclick-d3.tau-target-T.ci90.1\endcsname{$-$1.4}
\expandafter\def\csname val@rev.desc.d1.uiv.p1.Tsealed.zoomclick-d3.tau-target-T.ci95-excludes-0\endcsname{1}
\expandafter\def\csname cnt@rev.desc.d1.uiv.p1.Tsealed.zoomclick-d3.tau-target-T.ci95-excludes-0\endcsname{1}
\expandafter\def\csname pct@rev.desc.d1.uiv.p1.Tsealed.zoomclick-d3.tau-target-T.ci95-excludes-0\endcsname{100.0}
\expandafter\def\csname ppt@rev.desc.d1.uiv.p1.Tsealed.zoomclick-d3.tau-target-T.ci95-excludes-0\endcsname{$+$100.0}
\expandafter\def\csname val@rev.desc.d1.uiv.p1.Tsealed.zoomclick-d3.tau-target-T.ci95.0\endcsname{-0.111}
\expandafter\def\csname cnt@rev.desc.d1.uiv.p1.Tsealed.zoomclick-d3.tau-target-T.ci95.0\endcsname{0}
\expandafter\def\csname pct@rev.desc.d1.uiv.p1.Tsealed.zoomclick-d3.tau-target-T.ci95.0\endcsname{$-$11.1}
\expandafter\def\csname ppt@rev.desc.d1.uiv.p1.Tsealed.zoomclick-d3.tau-target-T.ci95.0\endcsname{$-$11.1}
\expandafter\def\csname val@rev.desc.d1.uiv.p1.Tsealed.zoomclick-d3.tau-target-T.ci95.1\endcsname{-0.007}
\expandafter\def\csname cnt@rev.desc.d1.uiv.p1.Tsealed.zoomclick-d3.tau-target-T.ci95.1\endcsname{0}
\expandafter\def\csname pct@rev.desc.d1.uiv.p1.Tsealed.zoomclick-d3.tau-target-T.ci95.1\endcsname{$-$0.7}
\expandafter\def\csname ppt@rev.desc.d1.uiv.p1.Tsealed.zoomclick-d3.tau-target-T.ci95.1\endcsname{$-$0.7}
\expandafter\def\csname val@rev.desc.d1.uiv.p1.Tsealed.zoomclick-d3.tau-target-T.mean\endcsname{-0.056}
\expandafter\def\csname cnt@rev.desc.d1.uiv.p1.Tsealed.zoomclick-d3.tau-target-T.mean\endcsname{0}
\expandafter\def\csname pct@rev.desc.d1.uiv.p1.Tsealed.zoomclick-d3.tau-target-T.mean\endcsname{$-$5.6}
\expandafter\def\csname ppt@rev.desc.d1.uiv.p1.Tsealed.zoomclick-d3.tau-target-T.mean\endcsname{$-$5.6}
\expandafter\def\csname val@rev.desc.d1.uiv.p1.Tsealed.zoomclick-d3.tau-target-T.n-apps\endcsname{24}
\expandafter\def\csname cnt@rev.desc.d1.uiv.p1.Tsealed.zoomclick-d3.tau-target-T.n-apps\endcsname{24}
\expandafter\def\csname pct@rev.desc.d1.uiv.p1.Tsealed.zoomclick-d3.tau-target-T.n-apps\endcsname{2400.0}
\expandafter\def\csname ppt@rev.desc.d1.uiv.p1.Tsealed.zoomclick-d3.tau-target-T.n-apps\endcsname{$+$2400.0}
\expandafter\def\csname val@rev.desc.d1.uiv.p1.Tsealed.zoomclick-d3.tau-target-T.tost1.ci90-inside\endcsname{0}
\expandafter\def\csname cnt@rev.desc.d1.uiv.p1.Tsealed.zoomclick-d3.tau-target-T.tost1.ci90-inside\endcsname{0}
\expandafter\def\csname pct@rev.desc.d1.uiv.p1.Tsealed.zoomclick-d3.tau-target-T.tost1.ci90-inside\endcsname{0.0}
\expandafter\def\csname ppt@rev.desc.d1.uiv.p1.Tsealed.zoomclick-d3.tau-target-T.tost1.ci90-inside\endcsname{0.0}
\expandafter\def\csname val@rev.desc.d1.uiv.p1.Tsealed.zoomclick-d3.tau-target-T.tost1.equivalent\endcsname{0}
\expandafter\def\csname cnt@rev.desc.d1.uiv.p1.Tsealed.zoomclick-d3.tau-target-T.tost1.equivalent\endcsname{0}
\expandafter\def\csname pct@rev.desc.d1.uiv.p1.Tsealed.zoomclick-d3.tau-target-T.tost1.equivalent\endcsname{0.0}
\expandafter\def\csname ppt@rev.desc.d1.uiv.p1.Tsealed.zoomclick-d3.tau-target-T.tost1.equivalent\endcsname{0.0}
\expandafter\def\csname val@rev.desc.d1.uiv.p1.Tsealed.zoomclick-d3.tau-target-T.tost1.p\endcsname{0.963}
\expandafter\def\csname cnt@rev.desc.d1.uiv.p1.Tsealed.zoomclick-d3.tau-target-T.tost1.p\endcsname{1}
\expandafter\def\csname pct@rev.desc.d1.uiv.p1.Tsealed.zoomclick-d3.tau-target-T.tost1.p\endcsname{96.3}
\expandafter\def\csname ppt@rev.desc.d1.uiv.p1.Tsealed.zoomclick-d3.tau-target-T.tost1.p\endcsname{$+$96.3}
\expandafter\def\csname val@rev.desc.d1.uiv.p1.Tsealed.zoomclick-d3.tau-target-T.tost5.ci90-inside\endcsname{0}
\expandafter\def\csname cnt@rev.desc.d1.uiv.p1.Tsealed.zoomclick-d3.tau-target-T.tost5.ci90-inside\endcsname{0}
\expandafter\def\csname pct@rev.desc.d1.uiv.p1.Tsealed.zoomclick-d3.tau-target-T.tost5.ci90-inside\endcsname{0.0}
\expandafter\def\csname ppt@rev.desc.d1.uiv.p1.Tsealed.zoomclick-d3.tau-target-T.tost5.ci90-inside\endcsname{0.0}
\expandafter\def\csname val@rev.desc.d1.uiv.p1.Tsealed.zoomclick-d3.tau-target-T.tost5.equivalent\endcsname{0}
\expandafter\def\csname cnt@rev.desc.d1.uiv.p1.Tsealed.zoomclick-d3.tau-target-T.tost5.equivalent\endcsname{0}
\expandafter\def\csname pct@rev.desc.d1.uiv.p1.Tsealed.zoomclick-d3.tau-target-T.tost5.equivalent\endcsname{0.0}
\expandafter\def\csname ppt@rev.desc.d1.uiv.p1.Tsealed.zoomclick-d3.tau-target-T.tost5.equivalent\endcsname{0.0}
\expandafter\def\csname val@rev.desc.d1.uiv.p1.Tsealed.zoomclick-d3.tau-target-T.tost5.p\endcsname{0.533}
\expandafter\def\csname cnt@rev.desc.d1.uiv.p1.Tsealed.zoomclick-d3.tau-target-T.tost5.p\endcsname{1}
\expandafter\def\csname pct@rev.desc.d1.uiv.p1.Tsealed.zoomclick-d3.tau-target-T.tost5.p\endcsname{53.3}
\expandafter\def\csname ppt@rev.desc.d1.uiv.p1.Tsealed.zoomclick-d3.tau-target-T.tost5.p\endcsname{$+$53.3}
\expandafter\def\csname val@rev.desc.d1.uiv.p1.Tsealed.zoomclick-d3.zero-estimate\endcsname{0}
\expandafter\def\csname cnt@rev.desc.d1.uiv.p1.Tsealed.zoomclick-d3.zero-estimate\endcsname{0}
\expandafter\def\csname pct@rev.desc.d1.uiv.p1.Tsealed.zoomclick-d3.zero-estimate\endcsname{0.0}
\expandafter\def\csname ppt@rev.desc.d1.uiv.p1.Tsealed.zoomclick-d3.zero-estimate\endcsname{0.0}
\expandafter\def\csname val@rev.desc.d1.uiv.p2.C.center-d1.delta-tau.ci90.0\endcsname{-0.062}
\expandafter\def\csname cnt@rev.desc.d1.uiv.p2.C.center-d1.delta-tau.ci90.0\endcsname{0}
\expandafter\def\csname pct@rev.desc.d1.uiv.p2.C.center-d1.delta-tau.ci90.0\endcsname{$-$6.2}
\expandafter\def\csname ppt@rev.desc.d1.uiv.p2.C.center-d1.delta-tau.ci90.0\endcsname{$-$6.2}
\expandafter\def\csname val@rev.desc.d1.uiv.p2.C.center-d1.delta-tau.ci90.1\endcsname{0.094}
\expandafter\def\csname cnt@rev.desc.d1.uiv.p2.C.center-d1.delta-tau.ci90.1\endcsname{0}
\expandafter\def\csname pct@rev.desc.d1.uiv.p2.C.center-d1.delta-tau.ci90.1\endcsname{9.4}
\expandafter\def\csname ppt@rev.desc.d1.uiv.p2.C.center-d1.delta-tau.ci90.1\endcsname{$+$9.4}
\expandafter\def\csname val@rev.desc.d1.uiv.p2.C.center-d1.delta-tau.ci95-excludes-0\endcsname{0}
\expandafter\def\csname cnt@rev.desc.d1.uiv.p2.C.center-d1.delta-tau.ci95-excludes-0\endcsname{0}
\expandafter\def\csname pct@rev.desc.d1.uiv.p2.C.center-d1.delta-tau.ci95-excludes-0\endcsname{0.0}
\expandafter\def\csname ppt@rev.desc.d1.uiv.p2.C.center-d1.delta-tau.ci95-excludes-0\endcsname{0.0}
\expandafter\def\csname val@rev.desc.d1.uiv.p2.C.center-d1.delta-tau.ci95.0\endcsname{-0.083}
\expandafter\def\csname cnt@rev.desc.d1.uiv.p2.C.center-d1.delta-tau.ci95.0\endcsname{0}
\expandafter\def\csname pct@rev.desc.d1.uiv.p2.C.center-d1.delta-tau.ci95.0\endcsname{$-$8.3}
\expandafter\def\csname ppt@rev.desc.d1.uiv.p2.C.center-d1.delta-tau.ci95.0\endcsname{$-$8.3}
\expandafter\def\csname val@rev.desc.d1.uiv.p2.C.center-d1.delta-tau.ci95.1\endcsname{0.115}
\expandafter\def\csname cnt@rev.desc.d1.uiv.p2.C.center-d1.delta-tau.ci95.1\endcsname{0}
\expandafter\def\csname pct@rev.desc.d1.uiv.p2.C.center-d1.delta-tau.ci95.1\endcsname{11.5}
\expandafter\def\csname ppt@rev.desc.d1.uiv.p2.C.center-d1.delta-tau.ci95.1\endcsname{$+$11.5}
\expandafter\def\csname val@rev.desc.d1.uiv.p2.C.center-d1.delta-tau.mean\endcsname{0.010}
\expandafter\def\csname cnt@rev.desc.d1.uiv.p2.C.center-d1.delta-tau.mean\endcsname{0}
\expandafter\def\csname pct@rev.desc.d1.uiv.p2.C.center-d1.delta-tau.mean\endcsname{1.0}
\expandafter\def\csname ppt@rev.desc.d1.uiv.p2.C.center-d1.delta-tau.mean\endcsname{$+$1.0}
\expandafter\def\csname val@rev.desc.d1.uiv.p2.C.center-d1.delta-tau.n-apps\endcsname{12}
\expandafter\def\csname cnt@rev.desc.d1.uiv.p2.C.center-d1.delta-tau.n-apps\endcsname{12}
\expandafter\def\csname pct@rev.desc.d1.uiv.p2.C.center-d1.delta-tau.n-apps\endcsname{1200.0}
\expandafter\def\csname ppt@rev.desc.d1.uiv.p2.C.center-d1.delta-tau.n-apps\endcsname{$+$1200.0}
\expandafter\def\csname val@rev.desc.d1.uiv.p2.C.center-d1.delta-tau.tost1.ci90-inside\endcsname{0}
\expandafter\def\csname cnt@rev.desc.d1.uiv.p2.C.center-d1.delta-tau.tost1.ci90-inside\endcsname{0}
\expandafter\def\csname pct@rev.desc.d1.uiv.p2.C.center-d1.delta-tau.tost1.ci90-inside\endcsname{0.0}
\expandafter\def\csname ppt@rev.desc.d1.uiv.p2.C.center-d1.delta-tau.tost1.ci90-inside\endcsname{0.0}
\expandafter\def\csname val@rev.desc.d1.uiv.p2.C.center-d1.delta-tau.tost1.equivalent\endcsname{0}
\expandafter\def\csname cnt@rev.desc.d1.uiv.p2.C.center-d1.delta-tau.tost1.equivalent\endcsname{0}
\expandafter\def\csname pct@rev.desc.d1.uiv.p2.C.center-d1.delta-tau.tost1.equivalent\endcsname{0.0}
\expandafter\def\csname ppt@rev.desc.d1.uiv.p2.C.center-d1.delta-tau.tost1.equivalent\endcsname{0.0}
\expandafter\def\csname val@rev.desc.d1.uiv.p2.C.center-d1.delta-tau.tost1.p\endcsname{0.548}
\expandafter\def\csname cnt@rev.desc.d1.uiv.p2.C.center-d1.delta-tau.tost1.p\endcsname{1}
\expandafter\def\csname pct@rev.desc.d1.uiv.p2.C.center-d1.delta-tau.tost1.p\endcsname{54.8}
\expandafter\def\csname ppt@rev.desc.d1.uiv.p2.C.center-d1.delta-tau.tost1.p\endcsname{$+$54.8}
\expandafter\def\csname val@rev.desc.d1.uiv.p2.C.center-d1.delta-tau.tost5.ci90-inside\endcsname{0}
\expandafter\def\csname cnt@rev.desc.d1.uiv.p2.C.center-d1.delta-tau.tost5.ci90-inside\endcsname{0}
\expandafter\def\csname pct@rev.desc.d1.uiv.p2.C.center-d1.delta-tau.tost5.ci90-inside\endcsname{0.0}
\expandafter\def\csname ppt@rev.desc.d1.uiv.p2.C.center-d1.delta-tau.tost5.ci90-inside\endcsname{0.0}
\expandafter\def\csname val@rev.desc.d1.uiv.p2.C.center-d1.delta-tau.tost5.equivalent\endcsname{0}
\expandafter\def\csname cnt@rev.desc.d1.uiv.p2.C.center-d1.delta-tau.tost5.equivalent\endcsname{0}
\expandafter\def\csname pct@rev.desc.d1.uiv.p2.C.center-d1.delta-tau.tost5.equivalent\endcsname{0.0}
\expandafter\def\csname ppt@rev.desc.d1.uiv.p2.C.center-d1.delta-tau.tost5.equivalent\endcsname{0.0}
\expandafter\def\csname val@rev.desc.d1.uiv.p2.C.center-d1.delta-tau.tost5.p\endcsname{0.237}
\expandafter\def\csname cnt@rev.desc.d1.uiv.p2.C.center-d1.delta-tau.tost5.p\endcsname{0}
\expandafter\def\csname pct@rev.desc.d1.uiv.p2.C.center-d1.delta-tau.tost5.p\endcsname{23.7}
\expandafter\def\csname ppt@rev.desc.d1.uiv.p2.C.center-d1.delta-tau.tost5.p\endcsname{$+$23.7}
\expandafter\def\csname val@rev.desc.d1.uiv.p2.C.center-d1.material-flip\endcsname{0}
\expandafter\def\csname cnt@rev.desc.d1.uiv.p2.C.center-d1.material-flip\endcsname{0}
\expandafter\def\csname pct@rev.desc.d1.uiv.p2.C.center-d1.material-flip\endcsname{0.0}
\expandafter\def\csname ppt@rev.desc.d1.uiv.p2.C.center-d1.material-flip\endcsname{0.0}
\expandafter\def\csname val@rev.desc.d1.uiv.p2.C.center-d1.n-apps\endcsname{12}
\expandafter\def\csname cnt@rev.desc.d1.uiv.p2.C.center-d1.n-apps\endcsname{12}
\expandafter\def\csname pct@rev.desc.d1.uiv.p2.C.center-d1.n-apps\endcsname{1200.0}
\expandafter\def\csname ppt@rev.desc.d1.uiv.p2.C.center-d1.n-apps\endcsname{$+$1200.0}
\expandafter\def\csname val@rev.desc.d1.uiv.p2.C.center-d1.n-tasks\endcsname{96}
\expandafter\def\csname cnt@rev.desc.d1.uiv.p2.C.center-d1.n-tasks\endcsname{96}
\expandafter\def\csname pct@rev.desc.d1.uiv.p2.C.center-d1.n-tasks\endcsname{9600.0}
\expandafter\def\csname ppt@rev.desc.d1.uiv.p2.C.center-d1.n-tasks\endcsname{$+$9600.0}
\expandafter\def\csname val@rev.desc.d1.uiv.p2.C.center-d1.sign-differs\endcsname{0}
\expandafter\def\csname cnt@rev.desc.d1.uiv.p2.C.center-d1.sign-differs\endcsname{0}
\expandafter\def\csname pct@rev.desc.d1.uiv.p2.C.center-d1.sign-differs\endcsname{0.0}
\expandafter\def\csname ppt@rev.desc.d1.uiv.p2.C.center-d1.sign-differs\endcsname{0.0}
\expandafter\def\csname val@rev.desc.d1.uiv.p2.C.center-d1.tau-source.ci90.0\endcsname{-0.062}
\expandafter\def\csname cnt@rev.desc.d1.uiv.p2.C.center-d1.tau-source.ci90.0\endcsname{0}
\expandafter\def\csname pct@rev.desc.d1.uiv.p2.C.center-d1.tau-source.ci90.0\endcsname{$-$6.2}
\expandafter\def\csname ppt@rev.desc.d1.uiv.p2.C.center-d1.tau-source.ci90.0\endcsname{$-$6.2}
\expandafter\def\csname val@rev.desc.d1.uiv.p2.C.center-d1.tau-source.ci90.1\endcsname{0.073}
\expandafter\def\csname cnt@rev.desc.d1.uiv.p2.C.center-d1.tau-source.ci90.1\endcsname{0}
\expandafter\def\csname pct@rev.desc.d1.uiv.p2.C.center-d1.tau-source.ci90.1\endcsname{7.3}
\expandafter\def\csname ppt@rev.desc.d1.uiv.p2.C.center-d1.tau-source.ci90.1\endcsname{$+$7.3}
\expandafter\def\csname val@rev.desc.d1.uiv.p2.C.center-d1.tau-source.ci95-excludes-0\endcsname{0}
\expandafter\def\csname cnt@rev.desc.d1.uiv.p2.C.center-d1.tau-source.ci95-excludes-0\endcsname{0}
\expandafter\def\csname pct@rev.desc.d1.uiv.p2.C.center-d1.tau-source.ci95-excludes-0\endcsname{0.0}
\expandafter\def\csname ppt@rev.desc.d1.uiv.p2.C.center-d1.tau-source.ci95-excludes-0\endcsname{0.0}
\expandafter\def\csname val@rev.desc.d1.uiv.p2.C.center-d1.tau-source.ci95.0\endcsname{-0.073}
\expandafter\def\csname cnt@rev.desc.d1.uiv.p2.C.center-d1.tau-source.ci95.0\endcsname{0}
\expandafter\def\csname pct@rev.desc.d1.uiv.p2.C.center-d1.tau-source.ci95.0\endcsname{$-$7.3}
\expandafter\def\csname ppt@rev.desc.d1.uiv.p2.C.center-d1.tau-source.ci95.0\endcsname{$-$7.3}
\expandafter\def\csname val@rev.desc.d1.uiv.p2.C.center-d1.tau-source.ci95.1\endcsname{0.083}
\expandafter\def\csname cnt@rev.desc.d1.uiv.p2.C.center-d1.tau-source.ci95.1\endcsname{0}
\expandafter\def\csname pct@rev.desc.d1.uiv.p2.C.center-d1.tau-source.ci95.1\endcsname{8.3}
\expandafter\def\csname ppt@rev.desc.d1.uiv.p2.C.center-d1.tau-source.ci95.1\endcsname{$+$8.3}
\expandafter\def\csname val@rev.desc.d1.uiv.p2.C.center-d1.tau-source.mean\endcsname{0.010}
\expandafter\def\csname cnt@rev.desc.d1.uiv.p2.C.center-d1.tau-source.mean\endcsname{0}
\expandafter\def\csname pct@rev.desc.d1.uiv.p2.C.center-d1.tau-source.mean\endcsname{1.0}
\expandafter\def\csname ppt@rev.desc.d1.uiv.p2.C.center-d1.tau-source.mean\endcsname{$+$1.0}
\expandafter\def\csname val@rev.desc.d1.uiv.p2.C.center-d1.tau-source.n-apps\endcsname{12}
\expandafter\def\csname cnt@rev.desc.d1.uiv.p2.C.center-d1.tau-source.n-apps\endcsname{12}
\expandafter\def\csname pct@rev.desc.d1.uiv.p2.C.center-d1.tau-source.n-apps\endcsname{1200.0}
\expandafter\def\csname ppt@rev.desc.d1.uiv.p2.C.center-d1.tau-source.n-apps\endcsname{$+$1200.0}
\expandafter\def\csname val@rev.desc.d1.uiv.p2.C.center-d1.tau-source.tost1.ci90-inside\endcsname{0}
\expandafter\def\csname cnt@rev.desc.d1.uiv.p2.C.center-d1.tau-source.tost1.ci90-inside\endcsname{0}
\expandafter\def\csname pct@rev.desc.d1.uiv.p2.C.center-d1.tau-source.tost1.ci90-inside\endcsname{0.0}
\expandafter\def\csname ppt@rev.desc.d1.uiv.p2.C.center-d1.tau-source.tost1.ci90-inside\endcsname{0.0}
\expandafter\def\csname val@rev.desc.d1.uiv.p2.C.center-d1.tau-source.tost1.equivalent\endcsname{0}
\expandafter\def\csname cnt@rev.desc.d1.uiv.p2.C.center-d1.tau-source.tost1.equivalent\endcsname{0}
\expandafter\def\csname pct@rev.desc.d1.uiv.p2.C.center-d1.tau-source.tost1.equivalent\endcsname{0.0}
\expandafter\def\csname ppt@rev.desc.d1.uiv.p2.C.center-d1.tau-source.tost1.equivalent\endcsname{0.0}
\expandafter\def\csname val@rev.desc.d1.uiv.p2.C.center-d1.tau-source.tost1.p\endcsname{0.549}
\expandafter\def\csname cnt@rev.desc.d1.uiv.p2.C.center-d1.tau-source.tost1.p\endcsname{1}
\expandafter\def\csname pct@rev.desc.d1.uiv.p2.C.center-d1.tau-source.tost1.p\endcsname{54.9}
\expandafter\def\csname ppt@rev.desc.d1.uiv.p2.C.center-d1.tau-source.tost1.p\endcsname{$+$54.9}
\expandafter\def\csname val@rev.desc.d1.uiv.p2.C.center-d1.tau-source.tost5.ci90-inside\endcsname{0}
\expandafter\def\csname cnt@rev.desc.d1.uiv.p2.C.center-d1.tau-source.tost5.ci90-inside\endcsname{0}
\expandafter\def\csname pct@rev.desc.d1.uiv.p2.C.center-d1.tau-source.tost5.ci90-inside\endcsname{0.0}
\expandafter\def\csname ppt@rev.desc.d1.uiv.p2.C.center-d1.tau-source.tost5.ci90-inside\endcsname{0.0}
\expandafter\def\csname val@rev.desc.d1.uiv.p2.C.center-d1.tau-source.tost5.equivalent\endcsname{0}
\expandafter\def\csname cnt@rev.desc.d1.uiv.p2.C.center-d1.tau-source.tost5.equivalent\endcsname{0}
\expandafter\def\csname pct@rev.desc.d1.uiv.p2.C.center-d1.tau-source.tost5.equivalent\endcsname{0.0}
\expandafter\def\csname ppt@rev.desc.d1.uiv.p2.C.center-d1.tau-source.tost5.equivalent\endcsname{0.0}
\expandafter\def\csname val@rev.desc.d1.uiv.p2.C.center-d1.tau-source.tost5.p\endcsname{0.185}
\expandafter\def\csname cnt@rev.desc.d1.uiv.p2.C.center-d1.tau-source.tost5.p\endcsname{0}
\expandafter\def\csname pct@rev.desc.d1.uiv.p2.C.center-d1.tau-source.tost5.p\endcsname{18.5}
\expandafter\def\csname ppt@rev.desc.d1.uiv.p2.C.center-d1.tau-source.tost5.p\endcsname{$+$18.5}
\expandafter\def\csname val@rev.desc.d1.uiv.p2.C.center-d1.tau-target.ci90.0\endcsname{-0.021}
\expandafter\def\csname cnt@rev.desc.d1.uiv.p2.C.center-d1.tau-target.ci90.0\endcsname{0}
\expandafter\def\csname pct@rev.desc.d1.uiv.p2.C.center-d1.tau-target.ci90.0\endcsname{$-$2.1}
\expandafter\def\csname ppt@rev.desc.d1.uiv.p2.C.center-d1.tau-target.ci90.0\endcsname{$-$2.1}
\expandafter\def\csname val@rev.desc.d1.uiv.p2.C.center-d1.tau-target.ci90.1\endcsname{0.073}
\expandafter\def\csname cnt@rev.desc.d1.uiv.p2.C.center-d1.tau-target.ci90.1\endcsname{0}
\expandafter\def\csname pct@rev.desc.d1.uiv.p2.C.center-d1.tau-target.ci90.1\endcsname{7.3}
\expandafter\def\csname ppt@rev.desc.d1.uiv.p2.C.center-d1.tau-target.ci90.1\endcsname{$+$7.3}
\expandafter\def\csname val@rev.desc.d1.uiv.p2.C.center-d1.tau-target.ci95-excludes-0\endcsname{0}
\expandafter\def\csname cnt@rev.desc.d1.uiv.p2.C.center-d1.tau-target.ci95-excludes-0\endcsname{0}
\expandafter\def\csname pct@rev.desc.d1.uiv.p2.C.center-d1.tau-target.ci95-excludes-0\endcsname{0.0}
\expandafter\def\csname ppt@rev.desc.d1.uiv.p2.C.center-d1.tau-target.ci95-excludes-0\endcsname{0.0}
\expandafter\def\csname val@rev.desc.d1.uiv.p2.C.center-d1.tau-target.ci95.0\endcsname{-0.031}
\expandafter\def\csname cnt@rev.desc.d1.uiv.p2.C.center-d1.tau-target.ci95.0\endcsname{0}
\expandafter\def\csname pct@rev.desc.d1.uiv.p2.C.center-d1.tau-target.ci95.0\endcsname{$-$3.1}
\expandafter\def\csname ppt@rev.desc.d1.uiv.p2.C.center-d1.tau-target.ci95.0\endcsname{$-$3.1}
\expandafter\def\csname val@rev.desc.d1.uiv.p2.C.center-d1.tau-target.ci95.1\endcsname{0.083}
\expandafter\def\csname cnt@rev.desc.d1.uiv.p2.C.center-d1.tau-target.ci95.1\endcsname{0}
\expandafter\def\csname pct@rev.desc.d1.uiv.p2.C.center-d1.tau-target.ci95.1\endcsname{8.3}
\expandafter\def\csname ppt@rev.desc.d1.uiv.p2.C.center-d1.tau-target.ci95.1\endcsname{$+$8.3}
\expandafter\def\csname val@rev.desc.d1.uiv.p2.C.center-d1.tau-target.mean\endcsname{0.021}
\expandafter\def\csname cnt@rev.desc.d1.uiv.p2.C.center-d1.tau-target.mean\endcsname{0}
\expandafter\def\csname pct@rev.desc.d1.uiv.p2.C.center-d1.tau-target.mean\endcsname{2.1}
\expandafter\def\csname ppt@rev.desc.d1.uiv.p2.C.center-d1.tau-target.mean\endcsname{$+$2.1}
\expandafter\def\csname val@rev.desc.d1.uiv.p2.C.center-d1.tau-target.n-apps\endcsname{12}
\expandafter\def\csname cnt@rev.desc.d1.uiv.p2.C.center-d1.tau-target.n-apps\endcsname{12}
\expandafter\def\csname pct@rev.desc.d1.uiv.p2.C.center-d1.tau-target.n-apps\endcsname{1200.0}
\expandafter\def\csname ppt@rev.desc.d1.uiv.p2.C.center-d1.tau-target.n-apps\endcsname{$+$1200.0}
\expandafter\def\csname val@rev.desc.d1.uiv.p2.C.center-d1.tau-target.tost1.ci90-inside\endcsname{0}
\expandafter\def\csname cnt@rev.desc.d1.uiv.p2.C.center-d1.tau-target.tost1.ci90-inside\endcsname{0}
\expandafter\def\csname pct@rev.desc.d1.uiv.p2.C.center-d1.tau-target.tost1.ci90-inside\endcsname{0.0}
\expandafter\def\csname ppt@rev.desc.d1.uiv.p2.C.center-d1.tau-target.tost1.ci90-inside\endcsname{0.0}
\expandafter\def\csname val@rev.desc.d1.uiv.p2.C.center-d1.tau-target.tost1.equivalent\endcsname{0}
\expandafter\def\csname cnt@rev.desc.d1.uiv.p2.C.center-d1.tau-target.tost1.equivalent\endcsname{0}
\expandafter\def\csname pct@rev.desc.d1.uiv.p2.C.center-d1.tau-target.tost1.equivalent\endcsname{0.0}
\expandafter\def\csname ppt@rev.desc.d1.uiv.p2.C.center-d1.tau-target.tost1.equivalent\endcsname{0.0}
\expandafter\def\csname val@rev.desc.d1.uiv.p2.C.center-d1.tau-target.tost1.p\endcsname{0.708}
\expandafter\def\csname cnt@rev.desc.d1.uiv.p2.C.center-d1.tau-target.tost1.p\endcsname{1}
\expandafter\def\csname pct@rev.desc.d1.uiv.p2.C.center-d1.tau-target.tost1.p\endcsname{70.8}
\expandafter\def\csname ppt@rev.desc.d1.uiv.p2.C.center-d1.tau-target.tost1.p\endcsname{$+$70.8}
\expandafter\def\csname val@rev.desc.d1.uiv.p2.C.center-d1.tau-target.tost5.ci90-inside\endcsname{0}
\expandafter\def\csname cnt@rev.desc.d1.uiv.p2.C.center-d1.tau-target.tost5.ci90-inside\endcsname{0}
\expandafter\def\csname pct@rev.desc.d1.uiv.p2.C.center-d1.tau-target.tost5.ci90-inside\endcsname{0.0}
\expandafter\def\csname ppt@rev.desc.d1.uiv.p2.C.center-d1.tau-target.tost5.ci90-inside\endcsname{0.0}
\expandafter\def\csname val@rev.desc.d1.uiv.p2.C.center-d1.tau-target.tost5.equivalent\endcsname{0}
\expandafter\def\csname cnt@rev.desc.d1.uiv.p2.C.center-d1.tau-target.tost5.equivalent\endcsname{0}
\expandafter\def\csname pct@rev.desc.d1.uiv.p2.C.center-d1.tau-target.tost5.equivalent\endcsname{0.0}
\expandafter\def\csname ppt@rev.desc.d1.uiv.p2.C.center-d1.tau-target.tost5.equivalent\endcsname{0.0}
\expandafter\def\csname val@rev.desc.d1.uiv.p2.C.center-d1.tau-target.tost5.p\endcsname{0.185}
\expandafter\def\csname cnt@rev.desc.d1.uiv.p2.C.center-d1.tau-target.tost5.p\endcsname{0}
\expandafter\def\csname pct@rev.desc.d1.uiv.p2.C.center-d1.tau-target.tost5.p\endcsname{18.5}
\expandafter\def\csname ppt@rev.desc.d1.uiv.p2.C.center-d1.tau-target.tost5.p\endcsname{$+$18.5}
\expandafter\def\csname val@rev.desc.d1.uiv.p2.C.center-d1.zero-estimate\endcsname{0}
\expandafter\def\csname cnt@rev.desc.d1.uiv.p2.C.center-d1.zero-estimate\endcsname{0}
\expandafter\def\csname pct@rev.desc.d1.uiv.p2.C.center-d1.zero-estimate\endcsname{0.0}
\expandafter\def\csname ppt@rev.desc.d1.uiv.p2.C.center-d1.zero-estimate\endcsname{0.0}
\expandafter\def\csname val@rev.desc.d1.uiv.p2.C.context-d1.delta-tau.ci90.0\endcsname{-0.062}
\expandafter\def\csname cnt@rev.desc.d1.uiv.p2.C.context-d1.delta-tau.ci90.0\endcsname{0}
\expandafter\def\csname pct@rev.desc.d1.uiv.p2.C.context-d1.delta-tau.ci90.0\endcsname{$-$6.2}
\expandafter\def\csname ppt@rev.desc.d1.uiv.p2.C.context-d1.delta-tau.ci90.0\endcsname{$-$6.2}
\expandafter\def\csname val@rev.desc.d1.uiv.p2.C.context-d1.delta-tau.ci90.1\endcsname{0.125}
\expandafter\def\csname cnt@rev.desc.d1.uiv.p2.C.context-d1.delta-tau.ci90.1\endcsname{0}
\expandafter\def\csname pct@rev.desc.d1.uiv.p2.C.context-d1.delta-tau.ci90.1\endcsname{12.5}
\expandafter\def\csname ppt@rev.desc.d1.uiv.p2.C.context-d1.delta-tau.ci90.1\endcsname{$+$12.5}
\expandafter\def\csname val@rev.desc.d1.uiv.p2.C.context-d1.delta-tau.ci95-excludes-0\endcsname{0}
\expandafter\def\csname cnt@rev.desc.d1.uiv.p2.C.context-d1.delta-tau.ci95-excludes-0\endcsname{0}
\expandafter\def\csname pct@rev.desc.d1.uiv.p2.C.context-d1.delta-tau.ci95-excludes-0\endcsname{0.0}
\expandafter\def\csname ppt@rev.desc.d1.uiv.p2.C.context-d1.delta-tau.ci95-excludes-0\endcsname{0.0}
\expandafter\def\csname val@rev.desc.d1.uiv.p2.C.context-d1.delta-tau.ci95.0\endcsname{-0.073}
\expandafter\def\csname cnt@rev.desc.d1.uiv.p2.C.context-d1.delta-tau.ci95.0\endcsname{0}
\expandafter\def\csname pct@rev.desc.d1.uiv.p2.C.context-d1.delta-tau.ci95.0\endcsname{$-$7.3}
\expandafter\def\csname ppt@rev.desc.d1.uiv.p2.C.context-d1.delta-tau.ci95.0\endcsname{$-$7.3}
\expandafter\def\csname val@rev.desc.d1.uiv.p2.C.context-d1.delta-tau.ci95.1\endcsname{0.146}
\expandafter\def\csname cnt@rev.desc.d1.uiv.p2.C.context-d1.delta-tau.ci95.1\endcsname{0}
\expandafter\def\csname pct@rev.desc.d1.uiv.p2.C.context-d1.delta-tau.ci95.1\endcsname{14.6}
\expandafter\def\csname ppt@rev.desc.d1.uiv.p2.C.context-d1.delta-tau.ci95.1\endcsname{$+$14.6}
\expandafter\def\csname val@rev.desc.d1.uiv.p2.C.context-d1.delta-tau.mean\endcsname{0.031}
\expandafter\def\csname cnt@rev.desc.d1.uiv.p2.C.context-d1.delta-tau.mean\endcsname{0}
\expandafter\def\csname pct@rev.desc.d1.uiv.p2.C.context-d1.delta-tau.mean\endcsname{3.1}
\expandafter\def\csname ppt@rev.desc.d1.uiv.p2.C.context-d1.delta-tau.mean\endcsname{$+$3.1}
\expandafter\def\csname val@rev.desc.d1.uiv.p2.C.context-d1.delta-tau.n-apps\endcsname{12}
\expandafter\def\csname cnt@rev.desc.d1.uiv.p2.C.context-d1.delta-tau.n-apps\endcsname{12}
\expandafter\def\csname pct@rev.desc.d1.uiv.p2.C.context-d1.delta-tau.n-apps\endcsname{1200.0}
\expandafter\def\csname ppt@rev.desc.d1.uiv.p2.C.context-d1.delta-tau.n-apps\endcsname{$+$1200.0}
\expandafter\def\csname val@rev.desc.d1.uiv.p2.C.context-d1.delta-tau.tost1.ci90-inside\endcsname{0}
\expandafter\def\csname cnt@rev.desc.d1.uiv.p2.C.context-d1.delta-tau.tost1.ci90-inside\endcsname{0}
\expandafter\def\csname pct@rev.desc.d1.uiv.p2.C.context-d1.delta-tau.tost1.ci90-inside\endcsname{0.0}
\expandafter\def\csname ppt@rev.desc.d1.uiv.p2.C.context-d1.delta-tau.tost1.ci90-inside\endcsname{0.0}
\expandafter\def\csname val@rev.desc.d1.uiv.p2.C.context-d1.delta-tau.tost1.equivalent\endcsname{0}
\expandafter\def\csname cnt@rev.desc.d1.uiv.p2.C.context-d1.delta-tau.tost1.equivalent\endcsname{0}
\expandafter\def\csname pct@rev.desc.d1.uiv.p2.C.context-d1.delta-tau.tost1.equivalent\endcsname{0.0}
\expandafter\def\csname ppt@rev.desc.d1.uiv.p2.C.context-d1.delta-tau.tost1.equivalent\endcsname{0.0}
\expandafter\def\csname val@rev.desc.d1.uiv.p2.C.context-d1.delta-tau.tost1.p\endcsname{0.667}
\expandafter\def\csname cnt@rev.desc.d1.uiv.p2.C.context-d1.delta-tau.tost1.p\endcsname{1}
\expandafter\def\csname pct@rev.desc.d1.uiv.p2.C.context-d1.delta-tau.tost1.p\endcsname{66.7}
\expandafter\def\csname ppt@rev.desc.d1.uiv.p2.C.context-d1.delta-tau.tost1.p\endcsname{$+$66.7}
\expandafter\def\csname val@rev.desc.d1.uiv.p2.C.context-d1.delta-tau.tost5.ci90-inside\endcsname{0}
\expandafter\def\csname cnt@rev.desc.d1.uiv.p2.C.context-d1.delta-tau.tost5.ci90-inside\endcsname{0}
\expandafter\def\csname pct@rev.desc.d1.uiv.p2.C.context-d1.delta-tau.tost5.ci90-inside\endcsname{0.0}
\expandafter\def\csname ppt@rev.desc.d1.uiv.p2.C.context-d1.delta-tau.tost5.ci90-inside\endcsname{0.0}
\expandafter\def\csname val@rev.desc.d1.uiv.p2.C.context-d1.delta-tau.tost5.equivalent\endcsname{0}
\expandafter\def\csname cnt@rev.desc.d1.uiv.p2.C.context-d1.delta-tau.tost5.equivalent\endcsname{0}
\expandafter\def\csname pct@rev.desc.d1.uiv.p2.C.context-d1.delta-tau.tost5.equivalent\endcsname{0.0}
\expandafter\def\csname ppt@rev.desc.d1.uiv.p2.C.context-d1.delta-tau.tost5.equivalent\endcsname{0.0}
\expandafter\def\csname val@rev.desc.d1.uiv.p2.C.context-d1.delta-tau.tost5.p\endcsname{0.376}
\expandafter\def\csname cnt@rev.desc.d1.uiv.p2.C.context-d1.delta-tau.tost5.p\endcsname{0}
\expandafter\def\csname pct@rev.desc.d1.uiv.p2.C.context-d1.delta-tau.tost5.p\endcsname{37.6}
\expandafter\def\csname ppt@rev.desc.d1.uiv.p2.C.context-d1.delta-tau.tost5.p\endcsname{$+$37.6}
\expandafter\def\csname val@rev.desc.d1.uiv.p2.C.context-d1.material-flip\endcsname{0}
\expandafter\def\csname cnt@rev.desc.d1.uiv.p2.C.context-d1.material-flip\endcsname{0}
\expandafter\def\csname pct@rev.desc.d1.uiv.p2.C.context-d1.material-flip\endcsname{0.0}
\expandafter\def\csname ppt@rev.desc.d1.uiv.p2.C.context-d1.material-flip\endcsname{0.0}
\expandafter\def\csname val@rev.desc.d1.uiv.p2.C.context-d1.n-apps\endcsname{12}
\expandafter\def\csname cnt@rev.desc.d1.uiv.p2.C.context-d1.n-apps\endcsname{12}
\expandafter\def\csname pct@rev.desc.d1.uiv.p2.C.context-d1.n-apps\endcsname{1200.0}
\expandafter\def\csname ppt@rev.desc.d1.uiv.p2.C.context-d1.n-apps\endcsname{$+$1200.0}
\expandafter\def\csname val@rev.desc.d1.uiv.p2.C.context-d1.n-tasks\endcsname{96}
\expandafter\def\csname cnt@rev.desc.d1.uiv.p2.C.context-d1.n-tasks\endcsname{96}
\expandafter\def\csname pct@rev.desc.d1.uiv.p2.C.context-d1.n-tasks\endcsname{9600.0}
\expandafter\def\csname ppt@rev.desc.d1.uiv.p2.C.context-d1.n-tasks\endcsname{$+$9600.0}
\expandafter\def\csname val@rev.desc.d1.uiv.p2.C.context-d1.sign-differs\endcsname{1}
\expandafter\def\csname cnt@rev.desc.d1.uiv.p2.C.context-d1.sign-differs\endcsname{1}
\expandafter\def\csname pct@rev.desc.d1.uiv.p2.C.context-d1.sign-differs\endcsname{100.0}
\expandafter\def\csname ppt@rev.desc.d1.uiv.p2.C.context-d1.sign-differs\endcsname{$+$100.0}
\expandafter\def\csname val@rev.desc.d1.uiv.p2.C.context-d1.tau-source.ci90.0\endcsname{-0.083}
\expandafter\def\csname cnt@rev.desc.d1.uiv.p2.C.context-d1.tau-source.ci90.0\endcsname{0}
\expandafter\def\csname pct@rev.desc.d1.uiv.p2.C.context-d1.tau-source.ci90.0\endcsname{$-$8.3}
\expandafter\def\csname ppt@rev.desc.d1.uiv.p2.C.context-d1.tau-source.ci90.0\endcsname{$-$8.3}
\expandafter\def\csname val@rev.desc.d1.uiv.p2.C.context-d1.tau-source.ci90.1\endcsname{0.042}
\expandafter\def\csname cnt@rev.desc.d1.uiv.p2.C.context-d1.tau-source.ci90.1\endcsname{0}
\expandafter\def\csname pct@rev.desc.d1.uiv.p2.C.context-d1.tau-source.ci90.1\endcsname{4.2}
\expandafter\def\csname ppt@rev.desc.d1.uiv.p2.C.context-d1.tau-source.ci90.1\endcsname{$+$4.2}
\expandafter\def\csname val@rev.desc.d1.uiv.p2.C.context-d1.tau-source.ci95-excludes-0\endcsname{0}
\expandafter\def\csname cnt@rev.desc.d1.uiv.p2.C.context-d1.tau-source.ci95-excludes-0\endcsname{0}
\expandafter\def\csname pct@rev.desc.d1.uiv.p2.C.context-d1.tau-source.ci95-excludes-0\endcsname{0.0}
\expandafter\def\csname ppt@rev.desc.d1.uiv.p2.C.context-d1.tau-source.ci95-excludes-0\endcsname{0.0}
\expandafter\def\csname val@rev.desc.d1.uiv.p2.C.context-d1.tau-source.ci95.0\endcsname{-0.104}
\expandafter\def\csname cnt@rev.desc.d1.uiv.p2.C.context-d1.tau-source.ci95.0\endcsname{0}
\expandafter\def\csname pct@rev.desc.d1.uiv.p2.C.context-d1.tau-source.ci95.0\endcsname{$-$10.4}
\expandafter\def\csname ppt@rev.desc.d1.uiv.p2.C.context-d1.tau-source.ci95.0\endcsname{$-$10.4}
\expandafter\def\csname val@rev.desc.d1.uiv.p2.C.context-d1.tau-source.ci95.1\endcsname{0.052}
\expandafter\def\csname cnt@rev.desc.d1.uiv.p2.C.context-d1.tau-source.ci95.1\endcsname{0}
\expandafter\def\csname pct@rev.desc.d1.uiv.p2.C.context-d1.tau-source.ci95.1\endcsname{5.2}
\expandafter\def\csname ppt@rev.desc.d1.uiv.p2.C.context-d1.tau-source.ci95.1\endcsname{$+$5.2}
\expandafter\def\csname val@rev.desc.d1.uiv.p2.C.context-d1.tau-source.mean\endcsname{-0.021}
\expandafter\def\csname cnt@rev.desc.d1.uiv.p2.C.context-d1.tau-source.mean\endcsname{0}
\expandafter\def\csname pct@rev.desc.d1.uiv.p2.C.context-d1.tau-source.mean\endcsname{$-$2.1}
\expandafter\def\csname ppt@rev.desc.d1.uiv.p2.C.context-d1.tau-source.mean\endcsname{$-$2.1}
\expandafter\def\csname val@rev.desc.d1.uiv.p2.C.context-d1.tau-source.n-apps\endcsname{12}
\expandafter\def\csname cnt@rev.desc.d1.uiv.p2.C.context-d1.tau-source.n-apps\endcsname{12}
\expandafter\def\csname pct@rev.desc.d1.uiv.p2.C.context-d1.tau-source.n-apps\endcsname{1200.0}
\expandafter\def\csname ppt@rev.desc.d1.uiv.p2.C.context-d1.tau-source.n-apps\endcsname{$+$1200.0}
\expandafter\def\csname val@rev.desc.d1.uiv.p2.C.context-d1.tau-source.tost1.ci90-inside\endcsname{0}
\expandafter\def\csname cnt@rev.desc.d1.uiv.p2.C.context-d1.tau-source.tost1.ci90-inside\endcsname{0}
\expandafter\def\csname pct@rev.desc.d1.uiv.p2.C.context-d1.tau-source.tost1.ci90-inside\endcsname{0.0}
\expandafter\def\csname ppt@rev.desc.d1.uiv.p2.C.context-d1.tau-source.tost1.ci90-inside\endcsname{0.0}
\expandafter\def\csname val@rev.desc.d1.uiv.p2.C.context-d1.tau-source.tost1.equivalent\endcsname{0}
\expandafter\def\csname cnt@rev.desc.d1.uiv.p2.C.context-d1.tau-source.tost1.equivalent\endcsname{0}
\expandafter\def\csname pct@rev.desc.d1.uiv.p2.C.context-d1.tau-source.tost1.equivalent\endcsname{0.0}
\expandafter\def\csname ppt@rev.desc.d1.uiv.p2.C.context-d1.tau-source.tost1.equivalent\endcsname{0.0}
\expandafter\def\csname val@rev.desc.d1.uiv.p2.C.context-d1.tau-source.tost1.p\endcsname{0.639}
\expandafter\def\csname cnt@rev.desc.d1.uiv.p2.C.context-d1.tau-source.tost1.p\endcsname{1}
\expandafter\def\csname pct@rev.desc.d1.uiv.p2.C.context-d1.tau-source.tost1.p\endcsname{63.9}
\expandafter\def\csname ppt@rev.desc.d1.uiv.p2.C.context-d1.tau-source.tost1.p\endcsname{$+$63.9}
\expandafter\def\csname val@rev.desc.d1.uiv.p2.C.context-d1.tau-source.tost5.ci90-inside\endcsname{0}
\expandafter\def\csname cnt@rev.desc.d1.uiv.p2.C.context-d1.tau-source.tost5.ci90-inside\endcsname{0}
\expandafter\def\csname pct@rev.desc.d1.uiv.p2.C.context-d1.tau-source.tost5.ci90-inside\endcsname{0.0}
\expandafter\def\csname ppt@rev.desc.d1.uiv.p2.C.context-d1.tau-source.tost5.ci90-inside\endcsname{0.0}
\expandafter\def\csname val@rev.desc.d1.uiv.p2.C.context-d1.tau-source.tost5.equivalent\endcsname{0}
\expandafter\def\csname cnt@rev.desc.d1.uiv.p2.C.context-d1.tau-source.tost5.equivalent\endcsname{0}
\expandafter\def\csname pct@rev.desc.d1.uiv.p2.C.context-d1.tau-source.tost5.equivalent\endcsname{0.0}
\expandafter\def\csname ppt@rev.desc.d1.uiv.p2.C.context-d1.tau-source.tost5.equivalent\endcsname{0.0}
\expandafter\def\csname val@rev.desc.d1.uiv.p2.C.context-d1.tau-source.tost5.p\endcsname{0.247}
\expandafter\def\csname cnt@rev.desc.d1.uiv.p2.C.context-d1.tau-source.tost5.p\endcsname{0}
\expandafter\def\csname pct@rev.desc.d1.uiv.p2.C.context-d1.tau-source.tost5.p\endcsname{24.7}
\expandafter\def\csname ppt@rev.desc.d1.uiv.p2.C.context-d1.tau-source.tost5.p\endcsname{$+$24.7}
\expandafter\def\csname val@rev.desc.d1.uiv.p2.C.context-d1.tau-target.ci90.0\endcsname{-0.073}
\expandafter\def\csname cnt@rev.desc.d1.uiv.p2.C.context-d1.tau-target.ci90.0\endcsname{0}
\expandafter\def\csname pct@rev.desc.d1.uiv.p2.C.context-d1.tau-target.ci90.0\endcsname{$-$7.3}
\expandafter\def\csname ppt@rev.desc.d1.uiv.p2.C.context-d1.tau-target.ci90.0\endcsname{$-$7.3}
\expandafter\def\csname val@rev.desc.d1.uiv.p2.C.context-d1.tau-target.ci90.1\endcsname{0.083}
\expandafter\def\csname cnt@rev.desc.d1.uiv.p2.C.context-d1.tau-target.ci90.1\endcsname{0}
\expandafter\def\csname pct@rev.desc.d1.uiv.p2.C.context-d1.tau-target.ci90.1\endcsname{8.3}
\expandafter\def\csname ppt@rev.desc.d1.uiv.p2.C.context-d1.tau-target.ci90.1\endcsname{$+$8.3}
\expandafter\def\csname val@rev.desc.d1.uiv.p2.C.context-d1.tau-target.ci95-excludes-0\endcsname{0}
\expandafter\def\csname cnt@rev.desc.d1.uiv.p2.C.context-d1.tau-target.ci95-excludes-0\endcsname{0}
\expandafter\def\csname pct@rev.desc.d1.uiv.p2.C.context-d1.tau-target.ci95-excludes-0\endcsname{0.0}
\expandafter\def\csname ppt@rev.desc.d1.uiv.p2.C.context-d1.tau-target.ci95-excludes-0\endcsname{0.0}
\expandafter\def\csname val@rev.desc.d1.uiv.p2.C.context-d1.tau-target.ci95.0\endcsname{-0.083}
\expandafter\def\csname cnt@rev.desc.d1.uiv.p2.C.context-d1.tau-target.ci95.0\endcsname{0}
\expandafter\def\csname pct@rev.desc.d1.uiv.p2.C.context-d1.tau-target.ci95.0\endcsname{$-$8.3}
\expandafter\def\csname ppt@rev.desc.d1.uiv.p2.C.context-d1.tau-target.ci95.0\endcsname{$-$8.3}
\expandafter\def\csname val@rev.desc.d1.uiv.p2.C.context-d1.tau-target.ci95.1\endcsname{0.104}
\expandafter\def\csname cnt@rev.desc.d1.uiv.p2.C.context-d1.tau-target.ci95.1\endcsname{0}
\expandafter\def\csname pct@rev.desc.d1.uiv.p2.C.context-d1.tau-target.ci95.1\endcsname{10.4}
\expandafter\def\csname ppt@rev.desc.d1.uiv.p2.C.context-d1.tau-target.ci95.1\endcsname{$+$10.4}
\expandafter\def\csname val@rev.desc.d1.uiv.p2.C.context-d1.tau-target.mean\endcsname{0.010}
\expandafter\def\csname cnt@rev.desc.d1.uiv.p2.C.context-d1.tau-target.mean\endcsname{0}
\expandafter\def\csname pct@rev.desc.d1.uiv.p2.C.context-d1.tau-target.mean\endcsname{1.0}
\expandafter\def\csname ppt@rev.desc.d1.uiv.p2.C.context-d1.tau-target.mean\endcsname{$+$1.0}
\expandafter\def\csname val@rev.desc.d1.uiv.p2.C.context-d1.tau-target.n-apps\endcsname{12}
\expandafter\def\csname cnt@rev.desc.d1.uiv.p2.C.context-d1.tau-target.n-apps\endcsname{12}
\expandafter\def\csname pct@rev.desc.d1.uiv.p2.C.context-d1.tau-target.n-apps\endcsname{1200.0}
\expandafter\def\csname ppt@rev.desc.d1.uiv.p2.C.context-d1.tau-target.n-apps\endcsname{$+$1200.0}
\expandafter\def\csname val@rev.desc.d1.uiv.p2.C.context-d1.tau-target.tost1.ci90-inside\endcsname{0}
\expandafter\def\csname cnt@rev.desc.d1.uiv.p2.C.context-d1.tau-target.tost1.ci90-inside\endcsname{0}
\expandafter\def\csname pct@rev.desc.d1.uiv.p2.C.context-d1.tau-target.tost1.ci90-inside\endcsname{0.0}
\expandafter\def\csname ppt@rev.desc.d1.uiv.p2.C.context-d1.tau-target.tost1.ci90-inside\endcsname{0.0}
\expandafter\def\csname val@rev.desc.d1.uiv.p2.C.context-d1.tau-target.tost1.equivalent\endcsname{0}
\expandafter\def\csname cnt@rev.desc.d1.uiv.p2.C.context-d1.tau-target.tost1.equivalent\endcsname{0}
\expandafter\def\csname pct@rev.desc.d1.uiv.p2.C.context-d1.tau-target.tost1.equivalent\endcsname{0.0}
\expandafter\def\csname ppt@rev.desc.d1.uiv.p2.C.context-d1.tau-target.tost1.equivalent\endcsname{0.0}
\expandafter\def\csname val@rev.desc.d1.uiv.p2.C.context-d1.tau-target.tost1.p\endcsname{0.544}
\expandafter\def\csname cnt@rev.desc.d1.uiv.p2.C.context-d1.tau-target.tost1.p\endcsname{1}
\expandafter\def\csname pct@rev.desc.d1.uiv.p2.C.context-d1.tau-target.tost1.p\endcsname{54.4}
\expandafter\def\csname ppt@rev.desc.d1.uiv.p2.C.context-d1.tau-target.tost1.p\endcsname{$+$54.4}
\expandafter\def\csname val@rev.desc.d1.uiv.p2.C.context-d1.tau-target.tost5.ci90-inside\endcsname{0}
\expandafter\def\csname cnt@rev.desc.d1.uiv.p2.C.context-d1.tau-target.tost5.ci90-inside\endcsname{0}
\expandafter\def\csname pct@rev.desc.d1.uiv.p2.C.context-d1.tau-target.tost5.ci90-inside\endcsname{0.0}
\expandafter\def\csname ppt@rev.desc.d1.uiv.p2.C.context-d1.tau-target.tost5.ci90-inside\endcsname{0.0}
\expandafter\def\csname val@rev.desc.d1.uiv.p2.C.context-d1.tau-target.tost5.equivalent\endcsname{0}
\expandafter\def\csname cnt@rev.desc.d1.uiv.p2.C.context-d1.tau-target.tost5.equivalent\endcsname{0}
\expandafter\def\csname pct@rev.desc.d1.uiv.p2.C.context-d1.tau-target.tost5.equivalent\endcsname{0.0}
\expandafter\def\csname ppt@rev.desc.d1.uiv.p2.C.context-d1.tau-target.tost5.equivalent\endcsname{0.0}
\expandafter\def\csname val@rev.desc.d1.uiv.p2.C.context-d1.tau-target.tost5.p\endcsname{0.221}
\expandafter\def\csname cnt@rev.desc.d1.uiv.p2.C.context-d1.tau-target.tost5.p\endcsname{0}
\expandafter\def\csname pct@rev.desc.d1.uiv.p2.C.context-d1.tau-target.tost5.p\endcsname{22.1}
\expandafter\def\csname ppt@rev.desc.d1.uiv.p2.C.context-d1.tau-target.tost5.p\endcsname{$+$22.1}
\expandafter\def\csname val@rev.desc.d1.uiv.p2.C.context-d1.zero-estimate\endcsname{0}
\expandafter\def\csname cnt@rev.desc.d1.uiv.p2.C.context-d1.zero-estimate\endcsname{0}
\expandafter\def\csname pct@rev.desc.d1.uiv.p2.C.context-d1.zero-estimate\endcsname{0.0}
\expandafter\def\csname ppt@rev.desc.d1.uiv.p2.C.context-d1.zero-estimate\endcsname{0.0}
\expandafter\def\csname val@rev.desc.d1.uiv.p2.C.dual-d2.delta-tau.ci90.0\endcsname{-0.083}
\expandafter\def\csname cnt@rev.desc.d1.uiv.p2.C.dual-d2.delta-tau.ci90.0\endcsname{0}
\expandafter\def\csname pct@rev.desc.d1.uiv.p2.C.dual-d2.delta-tau.ci90.0\endcsname{$-$8.3}
\expandafter\def\csname ppt@rev.desc.d1.uiv.p2.C.dual-d2.delta-tau.ci90.0\endcsname{$-$8.3}
\expandafter\def\csname val@rev.desc.d1.uiv.p2.C.dual-d2.delta-tau.ci90.1\endcsname{0.062}
\expandafter\def\csname cnt@rev.desc.d1.uiv.p2.C.dual-d2.delta-tau.ci90.1\endcsname{0}
\expandafter\def\csname pct@rev.desc.d1.uiv.p2.C.dual-d2.delta-tau.ci90.1\endcsname{6.2}
\expandafter\def\csname ppt@rev.desc.d1.uiv.p2.C.dual-d2.delta-tau.ci90.1\endcsname{$+$6.2}
\expandafter\def\csname val@rev.desc.d1.uiv.p2.C.dual-d2.delta-tau.ci95-excludes-0\endcsname{0}
\expandafter\def\csname cnt@rev.desc.d1.uiv.p2.C.dual-d2.delta-tau.ci95-excludes-0\endcsname{0}
\expandafter\def\csname pct@rev.desc.d1.uiv.p2.C.dual-d2.delta-tau.ci95-excludes-0\endcsname{0.0}
\expandafter\def\csname ppt@rev.desc.d1.uiv.p2.C.dual-d2.delta-tau.ci95-excludes-0\endcsname{0.0}
\expandafter\def\csname val@rev.desc.d1.uiv.p2.C.dual-d2.delta-tau.ci95.0\endcsname{-0.094}
\expandafter\def\csname cnt@rev.desc.d1.uiv.p2.C.dual-d2.delta-tau.ci95.0\endcsname{0}
\expandafter\def\csname pct@rev.desc.d1.uiv.p2.C.dual-d2.delta-tau.ci95.0\endcsname{$-$9.4}
\expandafter\def\csname ppt@rev.desc.d1.uiv.p2.C.dual-d2.delta-tau.ci95.0\endcsname{$-$9.4}
\expandafter\def\csname val@rev.desc.d1.uiv.p2.C.dual-d2.delta-tau.ci95.1\endcsname{0.073}
\expandafter\def\csname cnt@rev.desc.d1.uiv.p2.C.dual-d2.delta-tau.ci95.1\endcsname{0}
\expandafter\def\csname pct@rev.desc.d1.uiv.p2.C.dual-d2.delta-tau.ci95.1\endcsname{7.3}
\expandafter\def\csname ppt@rev.desc.d1.uiv.p2.C.dual-d2.delta-tau.ci95.1\endcsname{$+$7.3}
\expandafter\def\csname val@rev.desc.d1.uiv.p2.C.dual-d2.delta-tau.mean\endcsname{-0.010}
\expandafter\def\csname cnt@rev.desc.d1.uiv.p2.C.dual-d2.delta-tau.mean\endcsname{0}
\expandafter\def\csname pct@rev.desc.d1.uiv.p2.C.dual-d2.delta-tau.mean\endcsname{$-$1.0}
\expandafter\def\csname ppt@rev.desc.d1.uiv.p2.C.dual-d2.delta-tau.mean\endcsname{$-$1.0}
\expandafter\def\csname val@rev.desc.d1.uiv.p2.C.dual-d2.delta-tau.n-apps\endcsname{12}
\expandafter\def\csname cnt@rev.desc.d1.uiv.p2.C.dual-d2.delta-tau.n-apps\endcsname{12}
\expandafter\def\csname pct@rev.desc.d1.uiv.p2.C.dual-d2.delta-tau.n-apps\endcsname{1200.0}
\expandafter\def\csname ppt@rev.desc.d1.uiv.p2.C.dual-d2.delta-tau.n-apps\endcsname{$+$1200.0}
\expandafter\def\csname val@rev.desc.d1.uiv.p2.C.dual-d2.delta-tau.tost1.ci90-inside\endcsname{0}
\expandafter\def\csname cnt@rev.desc.d1.uiv.p2.C.dual-d2.delta-tau.tost1.ci90-inside\endcsname{0}
\expandafter\def\csname pct@rev.desc.d1.uiv.p2.C.dual-d2.delta-tau.tost1.ci90-inside\endcsname{0.0}
\expandafter\def\csname ppt@rev.desc.d1.uiv.p2.C.dual-d2.delta-tau.tost1.ci90-inside\endcsname{0.0}
\expandafter\def\csname val@rev.desc.d1.uiv.p2.C.dual-d2.delta-tau.tost1.equivalent\endcsname{0}
\expandafter\def\csname cnt@rev.desc.d1.uiv.p2.C.dual-d2.delta-tau.tost1.equivalent\endcsname{0}
\expandafter\def\csname pct@rev.desc.d1.uiv.p2.C.dual-d2.delta-tau.tost1.equivalent\endcsname{0.0}
\expandafter\def\csname ppt@rev.desc.d1.uiv.p2.C.dual-d2.delta-tau.tost1.equivalent\endcsname{0.0}
\expandafter\def\csname val@rev.desc.d1.uiv.p2.C.dual-d2.delta-tau.tost1.p\endcsname{0.554}
\expandafter\def\csname cnt@rev.desc.d1.uiv.p2.C.dual-d2.delta-tau.tost1.p\endcsname{1}
\expandafter\def\csname pct@rev.desc.d1.uiv.p2.C.dual-d2.delta-tau.tost1.p\endcsname{55.4}
\expandafter\def\csname ppt@rev.desc.d1.uiv.p2.C.dual-d2.delta-tau.tost1.p\endcsname{$+$55.4}
\expandafter\def\csname val@rev.desc.d1.uiv.p2.C.dual-d2.delta-tau.tost5.ci90-inside\endcsname{0}
\expandafter\def\csname cnt@rev.desc.d1.uiv.p2.C.dual-d2.delta-tau.tost5.ci90-inside\endcsname{0}
\expandafter\def\csname pct@rev.desc.d1.uiv.p2.C.dual-d2.delta-tau.tost5.ci90-inside\endcsname{0.0}
\expandafter\def\csname ppt@rev.desc.d1.uiv.p2.C.dual-d2.delta-tau.tost5.ci90-inside\endcsname{0.0}
\expandafter\def\csname val@rev.desc.d1.uiv.p2.C.dual-d2.delta-tau.tost5.equivalent\endcsname{0}
\expandafter\def\csname cnt@rev.desc.d1.uiv.p2.C.dual-d2.delta-tau.tost5.equivalent\endcsname{0}
\expandafter\def\csname pct@rev.desc.d1.uiv.p2.C.dual-d2.delta-tau.tost5.equivalent\endcsname{0.0}
\expandafter\def\csname ppt@rev.desc.d1.uiv.p2.C.dual-d2.delta-tau.tost5.equivalent\endcsname{0.0}
\expandafter\def\csname val@rev.desc.d1.uiv.p2.C.dual-d2.delta-tau.tost5.p\endcsname{0.201}
\expandafter\def\csname cnt@rev.desc.d1.uiv.p2.C.dual-d2.delta-tau.tost5.p\endcsname{0}
\expandafter\def\csname pct@rev.desc.d1.uiv.p2.C.dual-d2.delta-tau.tost5.p\endcsname{20.1}
\expandafter\def\csname ppt@rev.desc.d1.uiv.p2.C.dual-d2.delta-tau.tost5.p\endcsname{$+$20.1}
\expandafter\def\csname val@rev.desc.d1.uiv.p2.C.dual-d2.material-flip\endcsname{0}
\expandafter\def\csname cnt@rev.desc.d1.uiv.p2.C.dual-d2.material-flip\endcsname{0}
\expandafter\def\csname pct@rev.desc.d1.uiv.p2.C.dual-d2.material-flip\endcsname{0.0}
\expandafter\def\csname ppt@rev.desc.d1.uiv.p2.C.dual-d2.material-flip\endcsname{0.0}
\expandafter\def\csname val@rev.desc.d1.uiv.p2.C.dual-d2.n-apps\endcsname{12}
\expandafter\def\csname cnt@rev.desc.d1.uiv.p2.C.dual-d2.n-apps\endcsname{12}
\expandafter\def\csname pct@rev.desc.d1.uiv.p2.C.dual-d2.n-apps\endcsname{1200.0}
\expandafter\def\csname ppt@rev.desc.d1.uiv.p2.C.dual-d2.n-apps\endcsname{$+$1200.0}
\expandafter\def\csname val@rev.desc.d1.uiv.p2.C.dual-d2.n-tasks\endcsname{96}
\expandafter\def\csname cnt@rev.desc.d1.uiv.p2.C.dual-d2.n-tasks\endcsname{96}
\expandafter\def\csname pct@rev.desc.d1.uiv.p2.C.dual-d2.n-tasks\endcsname{9600.0}
\expandafter\def\csname ppt@rev.desc.d1.uiv.p2.C.dual-d2.n-tasks\endcsname{$+$9600.0}
\expandafter\def\csname val@rev.desc.d1.uiv.p2.C.dual-d2.sign-differs\endcsname{0}
\expandafter\def\csname cnt@rev.desc.d1.uiv.p2.C.dual-d2.sign-differs\endcsname{0}
\expandafter\def\csname pct@rev.desc.d1.uiv.p2.C.dual-d2.sign-differs\endcsname{0.0}
\expandafter\def\csname ppt@rev.desc.d1.uiv.p2.C.dual-d2.sign-differs\endcsname{0.0}
\expandafter\def\csname val@rev.desc.d1.uiv.p2.C.dual-d2.tau-source.ci90.0\endcsname{-0.073}
\expandafter\def\csname cnt@rev.desc.d1.uiv.p2.C.dual-d2.tau-source.ci90.0\endcsname{0}
\expandafter\def\csname pct@rev.desc.d1.uiv.p2.C.dual-d2.tau-source.ci90.0\endcsname{$-$7.3}
\expandafter\def\csname ppt@rev.desc.d1.uiv.p2.C.dual-d2.tau-source.ci90.0\endcsname{$-$7.3}
\expandafter\def\csname val@rev.desc.d1.uiv.p2.C.dual-d2.tau-source.ci90.1\endcsname{0.062}
\expandafter\def\csname cnt@rev.desc.d1.uiv.p2.C.dual-d2.tau-source.ci90.1\endcsname{0}
\expandafter\def\csname pct@rev.desc.d1.uiv.p2.C.dual-d2.tau-source.ci90.1\endcsname{6.2}
\expandafter\def\csname ppt@rev.desc.d1.uiv.p2.C.dual-d2.tau-source.ci90.1\endcsname{$+$6.2}
\expandafter\def\csname val@rev.desc.d1.uiv.p2.C.dual-d2.tau-source.ci95-excludes-0\endcsname{0}
\expandafter\def\csname cnt@rev.desc.d1.uiv.p2.C.dual-d2.tau-source.ci95-excludes-0\endcsname{0}
\expandafter\def\csname pct@rev.desc.d1.uiv.p2.C.dual-d2.tau-source.ci95-excludes-0\endcsname{0.0}
\expandafter\def\csname ppt@rev.desc.d1.uiv.p2.C.dual-d2.tau-source.ci95-excludes-0\endcsname{0.0}
\expandafter\def\csname val@rev.desc.d1.uiv.p2.C.dual-d2.tau-source.ci95.0\endcsname{-0.094}
\expandafter\def\csname cnt@rev.desc.d1.uiv.p2.C.dual-d2.tau-source.ci95.0\endcsname{0}
\expandafter\def\csname pct@rev.desc.d1.uiv.p2.C.dual-d2.tau-source.ci95.0\endcsname{$-$9.4}
\expandafter\def\csname ppt@rev.desc.d1.uiv.p2.C.dual-d2.tau-source.ci95.0\endcsname{$-$9.4}
\expandafter\def\csname val@rev.desc.d1.uiv.p2.C.dual-d2.tau-source.ci95.1\endcsname{0.073}
\expandafter\def\csname cnt@rev.desc.d1.uiv.p2.C.dual-d2.tau-source.ci95.1\endcsname{0}
\expandafter\def\csname pct@rev.desc.d1.uiv.p2.C.dual-d2.tau-source.ci95.1\endcsname{7.3}
\expandafter\def\csname ppt@rev.desc.d1.uiv.p2.C.dual-d2.tau-source.ci95.1\endcsname{$+$7.3}
\expandafter\def\csname val@rev.desc.d1.uiv.p2.C.dual-d2.tau-source.mean\endcsname{0.000}
\expandafter\def\csname cnt@rev.desc.d1.uiv.p2.C.dual-d2.tau-source.mean\endcsname{0}
\expandafter\def\csname pct@rev.desc.d1.uiv.p2.C.dual-d2.tau-source.mean\endcsname{0.0}
\expandafter\def\csname ppt@rev.desc.d1.uiv.p2.C.dual-d2.tau-source.mean\endcsname{0.0}
\expandafter\def\csname val@rev.desc.d1.uiv.p2.C.dual-d2.tau-source.n-apps\endcsname{12}
\expandafter\def\csname cnt@rev.desc.d1.uiv.p2.C.dual-d2.tau-source.n-apps\endcsname{12}
\expandafter\def\csname pct@rev.desc.d1.uiv.p2.C.dual-d2.tau-source.n-apps\endcsname{1200.0}
\expandafter\def\csname ppt@rev.desc.d1.uiv.p2.C.dual-d2.tau-source.n-apps\endcsname{$+$1200.0}
\expandafter\def\csname val@rev.desc.d1.uiv.p2.C.dual-d2.tau-source.tost1.ci90-inside\endcsname{0}
\expandafter\def\csname cnt@rev.desc.d1.uiv.p2.C.dual-d2.tau-source.tost1.ci90-inside\endcsname{0}
\expandafter\def\csname pct@rev.desc.d1.uiv.p2.C.dual-d2.tau-source.tost1.ci90-inside\endcsname{0.0}
\expandafter\def\csname ppt@rev.desc.d1.uiv.p2.C.dual-d2.tau-source.tost1.ci90-inside\endcsname{0.0}
\expandafter\def\csname val@rev.desc.d1.uiv.p2.C.dual-d2.tau-source.tost1.equivalent\endcsname{0}
\expandafter\def\csname cnt@rev.desc.d1.uiv.p2.C.dual-d2.tau-source.tost1.equivalent\endcsname{0}
\expandafter\def\csname pct@rev.desc.d1.uiv.p2.C.dual-d2.tau-source.tost1.equivalent\endcsname{0.0}
\expandafter\def\csname ppt@rev.desc.d1.uiv.p2.C.dual-d2.tau-source.tost1.equivalent\endcsname{0.0}
\expandafter\def\csname val@rev.desc.d1.uiv.p2.C.dual-d2.tau-source.tost1.p\endcsname{0.472}
\expandafter\def\csname cnt@rev.desc.d1.uiv.p2.C.dual-d2.tau-source.tost1.p\endcsname{0}
\expandafter\def\csname pct@rev.desc.d1.uiv.p2.C.dual-d2.tau-source.tost1.p\endcsname{47.2}
\expandafter\def\csname ppt@rev.desc.d1.uiv.p2.C.dual-d2.tau-source.tost1.p\endcsname{$+$47.2}
\expandafter\def\csname val@rev.desc.d1.uiv.p2.C.dual-d2.tau-source.tost5.ci90-inside\endcsname{0}
\expandafter\def\csname cnt@rev.desc.d1.uiv.p2.C.dual-d2.tau-source.tost5.ci90-inside\endcsname{0}
\expandafter\def\csname pct@rev.desc.d1.uiv.p2.C.dual-d2.tau-source.tost5.ci90-inside\endcsname{0.0}
\expandafter\def\csname ppt@rev.desc.d1.uiv.p2.C.dual-d2.tau-source.tost5.ci90-inside\endcsname{0.0}
\expandafter\def\csname val@rev.desc.d1.uiv.p2.C.dual-d2.tau-source.tost5.equivalent\endcsname{0}
\expandafter\def\csname cnt@rev.desc.d1.uiv.p2.C.dual-d2.tau-source.tost5.equivalent\endcsname{0}
\expandafter\def\csname pct@rev.desc.d1.uiv.p2.C.dual-d2.tau-source.tost5.equivalent\endcsname{0.0}
\expandafter\def\csname ppt@rev.desc.d1.uiv.p2.C.dual-d2.tau-source.tost5.equivalent\endcsname{0.0}
\expandafter\def\csname val@rev.desc.d1.uiv.p2.C.dual-d2.tau-source.tost5.p\endcsname{0.139}
\expandafter\def\csname cnt@rev.desc.d1.uiv.p2.C.dual-d2.tau-source.tost5.p\endcsname{0}
\expandafter\def\csname pct@rev.desc.d1.uiv.p2.C.dual-d2.tau-source.tost5.p\endcsname{13.9}
\expandafter\def\csname ppt@rev.desc.d1.uiv.p2.C.dual-d2.tau-source.tost5.p\endcsname{$+$13.9}
\expandafter\def\csname val@rev.desc.d1.uiv.p2.C.dual-d2.tau-target.ci90.0\endcsname{-0.073}
\expandafter\def\csname cnt@rev.desc.d1.uiv.p2.C.dual-d2.tau-target.ci90.0\endcsname{0}
\expandafter\def\csname pct@rev.desc.d1.uiv.p2.C.dual-d2.tau-target.ci90.0\endcsname{$-$7.3}
\expandafter\def\csname ppt@rev.desc.d1.uiv.p2.C.dual-d2.tau-target.ci90.0\endcsname{$-$7.3}
\expandafter\def\csname val@rev.desc.d1.uiv.p2.C.dual-d2.tau-target.ci90.1\endcsname{0.042}
\expandafter\def\csname cnt@rev.desc.d1.uiv.p2.C.dual-d2.tau-target.ci90.1\endcsname{0}
\expandafter\def\csname pct@rev.desc.d1.uiv.p2.C.dual-d2.tau-target.ci90.1\endcsname{4.2}
\expandafter\def\csname ppt@rev.desc.d1.uiv.p2.C.dual-d2.tau-target.ci90.1\endcsname{$+$4.2}
\expandafter\def\csname val@rev.desc.d1.uiv.p2.C.dual-d2.tau-target.ci95-excludes-0\endcsname{0}
\expandafter\def\csname cnt@rev.desc.d1.uiv.p2.C.dual-d2.tau-target.ci95-excludes-0\endcsname{0}
\expandafter\def\csname pct@rev.desc.d1.uiv.p2.C.dual-d2.tau-target.ci95-excludes-0\endcsname{0.0}
\expandafter\def\csname ppt@rev.desc.d1.uiv.p2.C.dual-d2.tau-target.ci95-excludes-0\endcsname{0.0}
\expandafter\def\csname val@rev.desc.d1.uiv.p2.C.dual-d2.tau-target.ci95.0\endcsname{-0.083}
\expandafter\def\csname cnt@rev.desc.d1.uiv.p2.C.dual-d2.tau-target.ci95.0\endcsname{0}
\expandafter\def\csname pct@rev.desc.d1.uiv.p2.C.dual-d2.tau-target.ci95.0\endcsname{$-$8.3}
\expandafter\def\csname ppt@rev.desc.d1.uiv.p2.C.dual-d2.tau-target.ci95.0\endcsname{$-$8.3}
\expandafter\def\csname val@rev.desc.d1.uiv.p2.C.dual-d2.tau-target.ci95.1\endcsname{0.052}
\expandafter\def\csname cnt@rev.desc.d1.uiv.p2.C.dual-d2.tau-target.ci95.1\endcsname{0}
\expandafter\def\csname pct@rev.desc.d1.uiv.p2.C.dual-d2.tau-target.ci95.1\endcsname{5.2}
\expandafter\def\csname ppt@rev.desc.d1.uiv.p2.C.dual-d2.tau-target.ci95.1\endcsname{$+$5.2}
\expandafter\def\csname val@rev.desc.d1.uiv.p2.C.dual-d2.tau-target.mean\endcsname{-0.010}
\expandafter\def\csname cnt@rev.desc.d1.uiv.p2.C.dual-d2.tau-target.mean\endcsname{0}
\expandafter\def\csname pct@rev.desc.d1.uiv.p2.C.dual-d2.tau-target.mean\endcsname{$-$1.0}
\expandafter\def\csname ppt@rev.desc.d1.uiv.p2.C.dual-d2.tau-target.mean\endcsname{$-$1.0}
\expandafter\def\csname val@rev.desc.d1.uiv.p2.C.dual-d2.tau-target.n-apps\endcsname{12}
\expandafter\def\csname cnt@rev.desc.d1.uiv.p2.C.dual-d2.tau-target.n-apps\endcsname{12}
\expandafter\def\csname pct@rev.desc.d1.uiv.p2.C.dual-d2.tau-target.n-apps\endcsname{1200.0}
\expandafter\def\csname ppt@rev.desc.d1.uiv.p2.C.dual-d2.tau-target.n-apps\endcsname{$+$1200.0}
\expandafter\def\csname val@rev.desc.d1.uiv.p2.C.dual-d2.tau-target.tost1.ci90-inside\endcsname{0}
\expandafter\def\csname cnt@rev.desc.d1.uiv.p2.C.dual-d2.tau-target.tost1.ci90-inside\endcsname{0}
\expandafter\def\csname pct@rev.desc.d1.uiv.p2.C.dual-d2.tau-target.tost1.ci90-inside\endcsname{0.0}
\expandafter\def\csname ppt@rev.desc.d1.uiv.p2.C.dual-d2.tau-target.tost1.ci90-inside\endcsname{0.0}
\expandafter\def\csname val@rev.desc.d1.uiv.p2.C.dual-d2.tau-target.tost1.equivalent\endcsname{0}
\expandafter\def\csname cnt@rev.desc.d1.uiv.p2.C.dual-d2.tau-target.tost1.equivalent\endcsname{0}
\expandafter\def\csname pct@rev.desc.d1.uiv.p2.C.dual-d2.tau-target.tost1.equivalent\endcsname{0.0}
\expandafter\def\csname ppt@rev.desc.d1.uiv.p2.C.dual-d2.tau-target.tost1.equivalent\endcsname{0.0}
\expandafter\def\csname val@rev.desc.d1.uiv.p2.C.dual-d2.tau-target.tost1.p\endcsname{0.543}
\expandafter\def\csname cnt@rev.desc.d1.uiv.p2.C.dual-d2.tau-target.tost1.p\endcsname{1}
\expandafter\def\csname pct@rev.desc.d1.uiv.p2.C.dual-d2.tau-target.tost1.p\endcsname{54.3}
\expandafter\def\csname ppt@rev.desc.d1.uiv.p2.C.dual-d2.tau-target.tost1.p\endcsname{$+$54.3}
\expandafter\def\csname val@rev.desc.d1.uiv.p2.C.dual-d2.tau-target.tost5.ci90-inside\endcsname{0}
\expandafter\def\csname cnt@rev.desc.d1.uiv.p2.C.dual-d2.tau-target.tost5.ci90-inside\endcsname{0}
\expandafter\def\csname pct@rev.desc.d1.uiv.p2.C.dual-d2.tau-target.tost5.ci90-inside\endcsname{0.0}
\expandafter\def\csname ppt@rev.desc.d1.uiv.p2.C.dual-d2.tau-target.tost5.ci90-inside\endcsname{0.0}
\expandafter\def\csname val@rev.desc.d1.uiv.p2.C.dual-d2.tau-target.tost5.equivalent\endcsname{0}
\expandafter\def\csname cnt@rev.desc.d1.uiv.p2.C.dual-d2.tau-target.tost5.equivalent\endcsname{0}
\expandafter\def\csname pct@rev.desc.d1.uiv.p2.C.dual-d2.tau-target.tost5.equivalent\endcsname{0.0}
\expandafter\def\csname ppt@rev.desc.d1.uiv.p2.C.dual-d2.tau-target.tost5.equivalent\endcsname{0.0}
\expandafter\def\csname val@rev.desc.d1.uiv.p2.C.dual-d2.tau-target.tost5.p\endcsname{0.145}
\expandafter\def\csname cnt@rev.desc.d1.uiv.p2.C.dual-d2.tau-target.tost5.p\endcsname{0}
\expandafter\def\csname pct@rev.desc.d1.uiv.p2.C.dual-d2.tau-target.tost5.p\endcsname{14.5}
\expandafter\def\csname ppt@rev.desc.d1.uiv.p2.C.dual-d2.tau-target.tost5.p\endcsname{$+$14.5}
\expandafter\def\csname val@rev.desc.d1.uiv.p2.C.dual-d2.zero-estimate\endcsname{1}
\expandafter\def\csname cnt@rev.desc.d1.uiv.p2.C.dual-d2.zero-estimate\endcsname{1}
\expandafter\def\csname pct@rev.desc.d1.uiv.p2.C.dual-d2.zero-estimate\endcsname{100.0}
\expandafter\def\csname ppt@rev.desc.d1.uiv.p2.C.dual-d2.zero-estimate\endcsname{$+$100.0}
\expandafter\def\csname val@rev.desc.d1.uiv.p2.C.grid2-d3.delta-tau.ci90.0\endcsname{-0.031}
\expandafter\def\csname cnt@rev.desc.d1.uiv.p2.C.grid2-d3.delta-tau.ci90.0\endcsname{0}
\expandafter\def\csname pct@rev.desc.d1.uiv.p2.C.grid2-d3.delta-tau.ci90.0\endcsname{$-$3.1}
\expandafter\def\csname ppt@rev.desc.d1.uiv.p2.C.grid2-d3.delta-tau.ci90.0\endcsname{$-$3.1}
\expandafter\def\csname val@rev.desc.d1.uiv.p2.C.grid2-d3.delta-tau.ci90.1\endcsname{0.125}
\expandafter\def\csname cnt@rev.desc.d1.uiv.p2.C.grid2-d3.delta-tau.ci90.1\endcsname{0}
\expandafter\def\csname pct@rev.desc.d1.uiv.p2.C.grid2-d3.delta-tau.ci90.1\endcsname{12.5}
\expandafter\def\csname ppt@rev.desc.d1.uiv.p2.C.grid2-d3.delta-tau.ci90.1\endcsname{$+$12.5}
\expandafter\def\csname val@rev.desc.d1.uiv.p2.C.grid2-d3.delta-tau.ci95-excludes-0\endcsname{0}
\expandafter\def\csname cnt@rev.desc.d1.uiv.p2.C.grid2-d3.delta-tau.ci95-excludes-0\endcsname{0}
\expandafter\def\csname pct@rev.desc.d1.uiv.p2.C.grid2-d3.delta-tau.ci95-excludes-0\endcsname{0.0}
\expandafter\def\csname ppt@rev.desc.d1.uiv.p2.C.grid2-d3.delta-tau.ci95-excludes-0\endcsname{0.0}
\expandafter\def\csname val@rev.desc.d1.uiv.p2.C.grid2-d3.delta-tau.ci95.0\endcsname{-0.042}
\expandafter\def\csname cnt@rev.desc.d1.uiv.p2.C.grid2-d3.delta-tau.ci95.0\endcsname{0}
\expandafter\def\csname pct@rev.desc.d1.uiv.p2.C.grid2-d3.delta-tau.ci95.0\endcsname{$-$4.2}
\expandafter\def\csname ppt@rev.desc.d1.uiv.p2.C.grid2-d3.delta-tau.ci95.0\endcsname{$-$4.2}
\expandafter\def\csname val@rev.desc.d1.uiv.p2.C.grid2-d3.delta-tau.ci95.1\endcsname{0.146}
\expandafter\def\csname cnt@rev.desc.d1.uiv.p2.C.grid2-d3.delta-tau.ci95.1\endcsname{0}
\expandafter\def\csname pct@rev.desc.d1.uiv.p2.C.grid2-d3.delta-tau.ci95.1\endcsname{14.6}
\expandafter\def\csname ppt@rev.desc.d1.uiv.p2.C.grid2-d3.delta-tau.ci95.1\endcsname{$+$14.6}
\expandafter\def\csname val@rev.desc.d1.uiv.p2.C.grid2-d3.delta-tau.mean\endcsname{0.042}
\expandafter\def\csname cnt@rev.desc.d1.uiv.p2.C.grid2-d3.delta-tau.mean\endcsname{0}
\expandafter\def\csname pct@rev.desc.d1.uiv.p2.C.grid2-d3.delta-tau.mean\endcsname{4.2}
\expandafter\def\csname ppt@rev.desc.d1.uiv.p2.C.grid2-d3.delta-tau.mean\endcsname{$+$4.2}
\expandafter\def\csname val@rev.desc.d1.uiv.p2.C.grid2-d3.delta-tau.n-apps\endcsname{12}
\expandafter\def\csname cnt@rev.desc.d1.uiv.p2.C.grid2-d3.delta-tau.n-apps\endcsname{12}
\expandafter\def\csname pct@rev.desc.d1.uiv.p2.C.grid2-d3.delta-tau.n-apps\endcsname{1200.0}
\expandafter\def\csname ppt@rev.desc.d1.uiv.p2.C.grid2-d3.delta-tau.n-apps\endcsname{$+$1200.0}
\expandafter\def\csname val@rev.desc.d1.uiv.p2.C.grid2-d3.delta-tau.tost1.ci90-inside\endcsname{0}
\expandafter\def\csname cnt@rev.desc.d1.uiv.p2.C.grid2-d3.delta-tau.tost1.ci90-inside\endcsname{0}
\expandafter\def\csname pct@rev.desc.d1.uiv.p2.C.grid2-d3.delta-tau.tost1.ci90-inside\endcsname{0.0}
\expandafter\def\csname ppt@rev.desc.d1.uiv.p2.C.grid2-d3.delta-tau.tost1.ci90-inside\endcsname{0.0}
\expandafter\def\csname val@rev.desc.d1.uiv.p2.C.grid2-d3.delta-tau.tost1.equivalent\endcsname{0}
\expandafter\def\csname cnt@rev.desc.d1.uiv.p2.C.grid2-d3.delta-tau.tost1.equivalent\endcsname{0}
\expandafter\def\csname pct@rev.desc.d1.uiv.p2.C.grid2-d3.delta-tau.tost1.equivalent\endcsname{0.0}
\expandafter\def\csname ppt@rev.desc.d1.uiv.p2.C.grid2-d3.delta-tau.tost1.equivalent\endcsname{0.0}
\expandafter\def\csname val@rev.desc.d1.uiv.p2.C.grid2-d3.delta-tau.tost1.p\endcsname{0.759}
\expandafter\def\csname cnt@rev.desc.d1.uiv.p2.C.grid2-d3.delta-tau.tost1.p\endcsname{1}
\expandafter\def\csname pct@rev.desc.d1.uiv.p2.C.grid2-d3.delta-tau.tost1.p\endcsname{75.9}
\expandafter\def\csname ppt@rev.desc.d1.uiv.p2.C.grid2-d3.delta-tau.tost1.p\endcsname{$+$75.9}
\expandafter\def\csname val@rev.desc.d1.uiv.p2.C.grid2-d3.delta-tau.tost5.ci90-inside\endcsname{0}
\expandafter\def\csname cnt@rev.desc.d1.uiv.p2.C.grid2-d3.delta-tau.tost5.ci90-inside\endcsname{0}
\expandafter\def\csname pct@rev.desc.d1.uiv.p2.C.grid2-d3.delta-tau.tost5.ci90-inside\endcsname{0.0}
\expandafter\def\csname ppt@rev.desc.d1.uiv.p2.C.grid2-d3.delta-tau.tost5.ci90-inside\endcsname{0.0}
\expandafter\def\csname val@rev.desc.d1.uiv.p2.C.grid2-d3.delta-tau.tost5.equivalent\endcsname{0}
\expandafter\def\csname cnt@rev.desc.d1.uiv.p2.C.grid2-d3.delta-tau.tost5.equivalent\endcsname{0}
\expandafter\def\csname pct@rev.desc.d1.uiv.p2.C.grid2-d3.delta-tau.tost5.equivalent\endcsname{0.0}
\expandafter\def\csname ppt@rev.desc.d1.uiv.p2.C.grid2-d3.delta-tau.tost5.equivalent\endcsname{0.0}
\expandafter\def\csname val@rev.desc.d1.uiv.p2.C.grid2-d3.delta-tau.tost5.p\endcsname{0.435}
\expandafter\def\csname cnt@rev.desc.d1.uiv.p2.C.grid2-d3.delta-tau.tost5.p\endcsname{0}
\expandafter\def\csname pct@rev.desc.d1.uiv.p2.C.grid2-d3.delta-tau.tost5.p\endcsname{43.5}
\expandafter\def\csname ppt@rev.desc.d1.uiv.p2.C.grid2-d3.delta-tau.tost5.p\endcsname{$+$43.5}
\expandafter\def\csname val@rev.desc.d1.uiv.p2.C.grid2-d3.material-flip\endcsname{0}
\expandafter\def\csname cnt@rev.desc.d1.uiv.p2.C.grid2-d3.material-flip\endcsname{0}
\expandafter\def\csname pct@rev.desc.d1.uiv.p2.C.grid2-d3.material-flip\endcsname{0.0}
\expandafter\def\csname ppt@rev.desc.d1.uiv.p2.C.grid2-d3.material-flip\endcsname{0.0}
\expandafter\def\csname val@rev.desc.d1.uiv.p2.C.grid2-d3.n-apps\endcsname{12}
\expandafter\def\csname cnt@rev.desc.d1.uiv.p2.C.grid2-d3.n-apps\endcsname{12}
\expandafter\def\csname pct@rev.desc.d1.uiv.p2.C.grid2-d3.n-apps\endcsname{1200.0}
\expandafter\def\csname ppt@rev.desc.d1.uiv.p2.C.grid2-d3.n-apps\endcsname{$+$1200.0}
\expandafter\def\csname val@rev.desc.d1.uiv.p2.C.grid2-d3.n-tasks\endcsname{96}
\expandafter\def\csname cnt@rev.desc.d1.uiv.p2.C.grid2-d3.n-tasks\endcsname{96}
\expandafter\def\csname pct@rev.desc.d1.uiv.p2.C.grid2-d3.n-tasks\endcsname{9600.0}
\expandafter\def\csname ppt@rev.desc.d1.uiv.p2.C.grid2-d3.n-tasks\endcsname{$+$9600.0}
\expandafter\def\csname val@rev.desc.d1.uiv.p2.C.grid2-d3.sign-differs\endcsname{1}
\expandafter\def\csname cnt@rev.desc.d1.uiv.p2.C.grid2-d3.sign-differs\endcsname{1}
\expandafter\def\csname pct@rev.desc.d1.uiv.p2.C.grid2-d3.sign-differs\endcsname{100.0}
\expandafter\def\csname ppt@rev.desc.d1.uiv.p2.C.grid2-d3.sign-differs\endcsname{$+$100.0}
\expandafter\def\csname val@rev.desc.d1.uiv.p2.C.grid2-d3.tau-source.ci90.0\endcsname{-0.094}
\expandafter\def\csname cnt@rev.desc.d1.uiv.p2.C.grid2-d3.tau-source.ci90.0\endcsname{0}
\expandafter\def\csname pct@rev.desc.d1.uiv.p2.C.grid2-d3.tau-source.ci90.0\endcsname{$-$9.4}
\expandafter\def\csname ppt@rev.desc.d1.uiv.p2.C.grid2-d3.tau-source.ci90.0\endcsname{$-$9.4}
\expandafter\def\csname val@rev.desc.d1.uiv.p2.C.grid2-d3.tau-source.ci90.1\endcsname{0.031}
\expandafter\def\csname cnt@rev.desc.d1.uiv.p2.C.grid2-d3.tau-source.ci90.1\endcsname{0}
\expandafter\def\csname pct@rev.desc.d1.uiv.p2.C.grid2-d3.tau-source.ci90.1\endcsname{3.1}
\expandafter\def\csname ppt@rev.desc.d1.uiv.p2.C.grid2-d3.tau-source.ci90.1\endcsname{$+$3.1}
\expandafter\def\csname val@rev.desc.d1.uiv.p2.C.grid2-d3.tau-source.ci95-excludes-0\endcsname{0}
\expandafter\def\csname cnt@rev.desc.d1.uiv.p2.C.grid2-d3.tau-source.ci95-excludes-0\endcsname{0}
\expandafter\def\csname pct@rev.desc.d1.uiv.p2.C.grid2-d3.tau-source.ci95-excludes-0\endcsname{0.0}
\expandafter\def\csname ppt@rev.desc.d1.uiv.p2.C.grid2-d3.tau-source.ci95-excludes-0\endcsname{0.0}
\expandafter\def\csname val@rev.desc.d1.uiv.p2.C.grid2-d3.tau-source.ci95.0\endcsname{-0.104}
\expandafter\def\csname cnt@rev.desc.d1.uiv.p2.C.grid2-d3.tau-source.ci95.0\endcsname{0}
\expandafter\def\csname pct@rev.desc.d1.uiv.p2.C.grid2-d3.tau-source.ci95.0\endcsname{$-$10.4}
\expandafter\def\csname ppt@rev.desc.d1.uiv.p2.C.grid2-d3.tau-source.ci95.0\endcsname{$-$10.4}
\expandafter\def\csname val@rev.desc.d1.uiv.p2.C.grid2-d3.tau-source.ci95.1\endcsname{0.042}
\expandafter\def\csname cnt@rev.desc.d1.uiv.p2.C.grid2-d3.tau-source.ci95.1\endcsname{0}
\expandafter\def\csname pct@rev.desc.d1.uiv.p2.C.grid2-d3.tau-source.ci95.1\endcsname{4.2}
\expandafter\def\csname ppt@rev.desc.d1.uiv.p2.C.grid2-d3.tau-source.ci95.1\endcsname{$+$4.2}
\expandafter\def\csname val@rev.desc.d1.uiv.p2.C.grid2-d3.tau-source.mean\endcsname{-0.031}
\expandafter\def\csname cnt@rev.desc.d1.uiv.p2.C.grid2-d3.tau-source.mean\endcsname{0}
\expandafter\def\csname pct@rev.desc.d1.uiv.p2.C.grid2-d3.tau-source.mean\endcsname{$-$3.1}
\expandafter\def\csname ppt@rev.desc.d1.uiv.p2.C.grid2-d3.tau-source.mean\endcsname{$-$3.1}
\expandafter\def\csname val@rev.desc.d1.uiv.p2.C.grid2-d3.tau-source.n-apps\endcsname{12}
\expandafter\def\csname cnt@rev.desc.d1.uiv.p2.C.grid2-d3.tau-source.n-apps\endcsname{12}
\expandafter\def\csname pct@rev.desc.d1.uiv.p2.C.grid2-d3.tau-source.n-apps\endcsname{1200.0}
\expandafter\def\csname ppt@rev.desc.d1.uiv.p2.C.grid2-d3.tau-source.n-apps\endcsname{$+$1200.0}
\expandafter\def\csname val@rev.desc.d1.uiv.p2.C.grid2-d3.tau-source.tost1.ci90-inside\endcsname{0}
\expandafter\def\csname cnt@rev.desc.d1.uiv.p2.C.grid2-d3.tau-source.tost1.ci90-inside\endcsname{0}
\expandafter\def\csname pct@rev.desc.d1.uiv.p2.C.grid2-d3.tau-source.tost1.ci90-inside\endcsname{0.0}
\expandafter\def\csname ppt@rev.desc.d1.uiv.p2.C.grid2-d3.tau-source.tost1.ci90-inside\endcsname{0.0}
\expandafter\def\csname val@rev.desc.d1.uiv.p2.C.grid2-d3.tau-source.tost1.equivalent\endcsname{0}
\expandafter\def\csname cnt@rev.desc.d1.uiv.p2.C.grid2-d3.tau-source.tost1.equivalent\endcsname{0}
\expandafter\def\csname pct@rev.desc.d1.uiv.p2.C.grid2-d3.tau-source.tost1.equivalent\endcsname{0.0}
\expandafter\def\csname ppt@rev.desc.d1.uiv.p2.C.grid2-d3.tau-source.tost1.equivalent\endcsname{0.0}
\expandafter\def\csname val@rev.desc.d1.uiv.p2.C.grid2-d3.tau-source.tost1.p\endcsname{0.757}
\expandafter\def\csname cnt@rev.desc.d1.uiv.p2.C.grid2-d3.tau-source.tost1.p\endcsname{1}
\expandafter\def\csname pct@rev.desc.d1.uiv.p2.C.grid2-d3.tau-source.tost1.p\endcsname{75.7}
\expandafter\def\csname ppt@rev.desc.d1.uiv.p2.C.grid2-d3.tau-source.tost1.p\endcsname{$+$75.7}
\expandafter\def\csname val@rev.desc.d1.uiv.p2.C.grid2-d3.tau-source.tost5.ci90-inside\endcsname{0}
\expandafter\def\csname cnt@rev.desc.d1.uiv.p2.C.grid2-d3.tau-source.tost5.ci90-inside\endcsname{0}
\expandafter\def\csname pct@rev.desc.d1.uiv.p2.C.grid2-d3.tau-source.tost5.ci90-inside\endcsname{0.0}
\expandafter\def\csname ppt@rev.desc.d1.uiv.p2.C.grid2-d3.tau-source.tost5.ci90-inside\endcsname{0.0}
\expandafter\def\csname val@rev.desc.d1.uiv.p2.C.grid2-d3.tau-source.tost5.equivalent\endcsname{0}
\expandafter\def\csname cnt@rev.desc.d1.uiv.p2.C.grid2-d3.tau-source.tost5.equivalent\endcsname{0}
\expandafter\def\csname pct@rev.desc.d1.uiv.p2.C.grid2-d3.tau-source.tost5.equivalent\endcsname{0.0}
\expandafter\def\csname ppt@rev.desc.d1.uiv.p2.C.grid2-d3.tau-source.tost5.equivalent\endcsname{0.0}
\expandafter\def\csname val@rev.desc.d1.uiv.p2.C.grid2-d3.tau-source.tost5.p\endcsname{0.334}
\expandafter\def\csname cnt@rev.desc.d1.uiv.p2.C.grid2-d3.tau-source.tost5.p\endcsname{0}
\expandafter\def\csname pct@rev.desc.d1.uiv.p2.C.grid2-d3.tau-source.tost5.p\endcsname{33.4}
\expandafter\def\csname ppt@rev.desc.d1.uiv.p2.C.grid2-d3.tau-source.tost5.p\endcsname{$+$33.4}
\expandafter\def\csname val@rev.desc.d1.uiv.p2.C.grid2-d3.tau-target.ci90.0\endcsname{-0.042}
\expandafter\def\csname cnt@rev.desc.d1.uiv.p2.C.grid2-d3.tau-target.ci90.0\endcsname{0}
\expandafter\def\csname pct@rev.desc.d1.uiv.p2.C.grid2-d3.tau-target.ci90.0\endcsname{$-$4.2}
\expandafter\def\csname ppt@rev.desc.d1.uiv.p2.C.grid2-d3.tau-target.ci90.0\endcsname{$-$4.2}
\expandafter\def\csname val@rev.desc.d1.uiv.p2.C.grid2-d3.tau-target.ci90.1\endcsname{0.062}
\expandafter\def\csname cnt@rev.desc.d1.uiv.p2.C.grid2-d3.tau-target.ci90.1\endcsname{0}
\expandafter\def\csname pct@rev.desc.d1.uiv.p2.C.grid2-d3.tau-target.ci90.1\endcsname{6.2}
\expandafter\def\csname ppt@rev.desc.d1.uiv.p2.C.grid2-d3.tau-target.ci90.1\endcsname{$+$6.2}
\expandafter\def\csname val@rev.desc.d1.uiv.p2.C.grid2-d3.tau-target.ci95-excludes-0\endcsname{0}
\expandafter\def\csname cnt@rev.desc.d1.uiv.p2.C.grid2-d3.tau-target.ci95-excludes-0\endcsname{0}
\expandafter\def\csname pct@rev.desc.d1.uiv.p2.C.grid2-d3.tau-target.ci95-excludes-0\endcsname{0.0}
\expandafter\def\csname ppt@rev.desc.d1.uiv.p2.C.grid2-d3.tau-target.ci95-excludes-0\endcsname{0.0}
\expandafter\def\csname val@rev.desc.d1.uiv.p2.C.grid2-d3.tau-target.ci95.0\endcsname{-0.052}
\expandafter\def\csname cnt@rev.desc.d1.uiv.p2.C.grid2-d3.tau-target.ci95.0\endcsname{0}
\expandafter\def\csname pct@rev.desc.d1.uiv.p2.C.grid2-d3.tau-target.ci95.0\endcsname{$-$5.2}
\expandafter\def\csname ppt@rev.desc.d1.uiv.p2.C.grid2-d3.tau-target.ci95.0\endcsname{$-$5.2}
\expandafter\def\csname val@rev.desc.d1.uiv.p2.C.grid2-d3.tau-target.ci95.1\endcsname{0.073}
\expandafter\def\csname cnt@rev.desc.d1.uiv.p2.C.grid2-d3.tau-target.ci95.1\endcsname{0}
\expandafter\def\csname pct@rev.desc.d1.uiv.p2.C.grid2-d3.tau-target.ci95.1\endcsname{7.3}
\expandafter\def\csname ppt@rev.desc.d1.uiv.p2.C.grid2-d3.tau-target.ci95.1\endcsname{$+$7.3}
\expandafter\def\csname val@rev.desc.d1.uiv.p2.C.grid2-d3.tau-target.mean\endcsname{0.010}
\expandafter\def\csname cnt@rev.desc.d1.uiv.p2.C.grid2-d3.tau-target.mean\endcsname{0}
\expandafter\def\csname pct@rev.desc.d1.uiv.p2.C.grid2-d3.tau-target.mean\endcsname{1.0}
\expandafter\def\csname ppt@rev.desc.d1.uiv.p2.C.grid2-d3.tau-target.mean\endcsname{$+$1.0}
\expandafter\def\csname val@rev.desc.d1.uiv.p2.C.grid2-d3.tau-target.n-apps\endcsname{12}
\expandafter\def\csname cnt@rev.desc.d1.uiv.p2.C.grid2-d3.tau-target.n-apps\endcsname{12}
\expandafter\def\csname pct@rev.desc.d1.uiv.p2.C.grid2-d3.tau-target.n-apps\endcsname{1200.0}
\expandafter\def\csname ppt@rev.desc.d1.uiv.p2.C.grid2-d3.tau-target.n-apps\endcsname{$+$1200.0}
\expandafter\def\csname val@rev.desc.d1.uiv.p2.C.grid2-d3.tau-target.tost1.ci90-inside\endcsname{0}
\expandafter\def\csname cnt@rev.desc.d1.uiv.p2.C.grid2-d3.tau-target.tost1.ci90-inside\endcsname{0}
\expandafter\def\csname pct@rev.desc.d1.uiv.p2.C.grid2-d3.tau-target.tost1.ci90-inside\endcsname{0.0}
\expandafter\def\csname ppt@rev.desc.d1.uiv.p2.C.grid2-d3.tau-target.tost1.ci90-inside\endcsname{0.0}
\expandafter\def\csname val@rev.desc.d1.uiv.p2.C.grid2-d3.tau-target.tost1.equivalent\endcsname{0}
\expandafter\def\csname cnt@rev.desc.d1.uiv.p2.C.grid2-d3.tau-target.tost1.equivalent\endcsname{0}
\expandafter\def\csname pct@rev.desc.d1.uiv.p2.C.grid2-d3.tau-target.tost1.equivalent\endcsname{0.0}
\expandafter\def\csname ppt@rev.desc.d1.uiv.p2.C.grid2-d3.tau-target.tost1.equivalent\endcsname{0.0}
\expandafter\def\csname val@rev.desc.d1.uiv.p2.C.grid2-d3.tau-target.tost1.p\endcsname{0.557}
\expandafter\def\csname cnt@rev.desc.d1.uiv.p2.C.grid2-d3.tau-target.tost1.p\endcsname{1}
\expandafter\def\csname pct@rev.desc.d1.uiv.p2.C.grid2-d3.tau-target.tost1.p\endcsname{55.7}
\expandafter\def\csname ppt@rev.desc.d1.uiv.p2.C.grid2-d3.tau-target.tost1.p\endcsname{$+$55.7}
\expandafter\def\csname val@rev.desc.d1.uiv.p2.C.grid2-d3.tau-target.tost5.ci90-inside\endcsname{0}
\expandafter\def\csname cnt@rev.desc.d1.uiv.p2.C.grid2-d3.tau-target.tost5.ci90-inside\endcsname{0}
\expandafter\def\csname pct@rev.desc.d1.uiv.p2.C.grid2-d3.tau-target.tost5.ci90-inside\endcsname{0.0}
\expandafter\def\csname ppt@rev.desc.d1.uiv.p2.C.grid2-d3.tau-target.tost5.ci90-inside\endcsname{0.0}
\expandafter\def\csname val@rev.desc.d1.uiv.p2.C.grid2-d3.tau-target.tost5.equivalent\endcsname{0}
\expandafter\def\csname cnt@rev.desc.d1.uiv.p2.C.grid2-d3.tau-target.tost5.equivalent\endcsname{0}
\expandafter\def\csname pct@rev.desc.d1.uiv.p2.C.grid2-d3.tau-target.tost5.equivalent\endcsname{0.0}
\expandafter\def\csname ppt@rev.desc.d1.uiv.p2.C.grid2-d3.tau-target.tost5.equivalent\endcsname{0.0}
\expandafter\def\csname val@rev.desc.d1.uiv.p2.C.grid2-d3.tau-target.tost5.p\endcsname{0.117}
\expandafter\def\csname cnt@rev.desc.d1.uiv.p2.C.grid2-d3.tau-target.tost5.p\endcsname{0}
\expandafter\def\csname pct@rev.desc.d1.uiv.p2.C.grid2-d3.tau-target.tost5.p\endcsname{11.7}
\expandafter\def\csname ppt@rev.desc.d1.uiv.p2.C.grid2-d3.tau-target.tost5.p\endcsname{$+$11.7}
\expandafter\def\csname val@rev.desc.d1.uiv.p2.C.grid2-d3.zero-estimate\endcsname{0}
\expandafter\def\csname cnt@rev.desc.d1.uiv.p2.C.grid2-d3.zero-estimate\endcsname{0}
\expandafter\def\csname pct@rev.desc.d1.uiv.p2.C.grid2-d3.zero-estimate\endcsname{0.0}
\expandafter\def\csname ppt@rev.desc.d1.uiv.p2.C.grid2-d3.zero-estimate\endcsname{0.0}
\expandafter\def\csname val@rev.desc.d1.uiv.p2.C.zoomclick-d3.delta-tau.ci90.0\endcsname{-0.073}
\expandafter\def\csname cnt@rev.desc.d1.uiv.p2.C.zoomclick-d3.delta-tau.ci90.0\endcsname{0}
\expandafter\def\csname pct@rev.desc.d1.uiv.p2.C.zoomclick-d3.delta-tau.ci90.0\endcsname{$-$7.3}
\expandafter\def\csname ppt@rev.desc.d1.uiv.p2.C.zoomclick-d3.delta-tau.ci90.0\endcsname{$-$7.3}
\expandafter\def\csname val@rev.desc.d1.uiv.p2.C.zoomclick-d3.delta-tau.ci90.1\endcsname{0.135}
\expandafter\def\csname cnt@rev.desc.d1.uiv.p2.C.zoomclick-d3.delta-tau.ci90.1\endcsname{0}
\expandafter\def\csname pct@rev.desc.d1.uiv.p2.C.zoomclick-d3.delta-tau.ci90.1\endcsname{13.5}
\expandafter\def\csname ppt@rev.desc.d1.uiv.p2.C.zoomclick-d3.delta-tau.ci90.1\endcsname{$+$13.5}
\expandafter\def\csname val@rev.desc.d1.uiv.p2.C.zoomclick-d3.delta-tau.ci95-excludes-0\endcsname{0}
\expandafter\def\csname cnt@rev.desc.d1.uiv.p2.C.zoomclick-d3.delta-tau.ci95-excludes-0\endcsname{0}
\expandafter\def\csname pct@rev.desc.d1.uiv.p2.C.zoomclick-d3.delta-tau.ci95-excludes-0\endcsname{0.0}
\expandafter\def\csname ppt@rev.desc.d1.uiv.p2.C.zoomclick-d3.delta-tau.ci95-excludes-0\endcsname{0.0}
\expandafter\def\csname val@rev.desc.d1.uiv.p2.C.zoomclick-d3.delta-tau.ci95.0\endcsname{-0.083}
\expandafter\def\csname cnt@rev.desc.d1.uiv.p2.C.zoomclick-d3.delta-tau.ci95.0\endcsname{0}
\expandafter\def\csname pct@rev.desc.d1.uiv.p2.C.zoomclick-d3.delta-tau.ci95.0\endcsname{$-$8.3}
\expandafter\def\csname ppt@rev.desc.d1.uiv.p2.C.zoomclick-d3.delta-tau.ci95.0\endcsname{$-$8.3}
\expandafter\def\csname val@rev.desc.d1.uiv.p2.C.zoomclick-d3.delta-tau.ci95.1\endcsname{0.156}
\expandafter\def\csname cnt@rev.desc.d1.uiv.p2.C.zoomclick-d3.delta-tau.ci95.1\endcsname{0}
\expandafter\def\csname pct@rev.desc.d1.uiv.p2.C.zoomclick-d3.delta-tau.ci95.1\endcsname{15.6}
\expandafter\def\csname ppt@rev.desc.d1.uiv.p2.C.zoomclick-d3.delta-tau.ci95.1\endcsname{$+$15.6}
\expandafter\def\csname val@rev.desc.d1.uiv.p2.C.zoomclick-d3.delta-tau.mean\endcsname{0.031}
\expandafter\def\csname cnt@rev.desc.d1.uiv.p2.C.zoomclick-d3.delta-tau.mean\endcsname{0}
\expandafter\def\csname pct@rev.desc.d1.uiv.p2.C.zoomclick-d3.delta-tau.mean\endcsname{3.1}
\expandafter\def\csname ppt@rev.desc.d1.uiv.p2.C.zoomclick-d3.delta-tau.mean\endcsname{$+$3.1}
\expandafter\def\csname val@rev.desc.d1.uiv.p2.C.zoomclick-d3.delta-tau.n-apps\endcsname{12}
\expandafter\def\csname cnt@rev.desc.d1.uiv.p2.C.zoomclick-d3.delta-tau.n-apps\endcsname{12}
\expandafter\def\csname pct@rev.desc.d1.uiv.p2.C.zoomclick-d3.delta-tau.n-apps\endcsname{1200.0}
\expandafter\def\csname ppt@rev.desc.d1.uiv.p2.C.zoomclick-d3.delta-tau.n-apps\endcsname{$+$1200.0}
\expandafter\def\csname val@rev.desc.d1.uiv.p2.C.zoomclick-d3.delta-tau.tost1.ci90-inside\endcsname{0}
\expandafter\def\csname cnt@rev.desc.d1.uiv.p2.C.zoomclick-d3.delta-tau.tost1.ci90-inside\endcsname{0}
\expandafter\def\csname pct@rev.desc.d1.uiv.p2.C.zoomclick-d3.delta-tau.tost1.ci90-inside\endcsname{0.0}
\expandafter\def\csname ppt@rev.desc.d1.uiv.p2.C.zoomclick-d3.delta-tau.tost1.ci90-inside\endcsname{0.0}
\expandafter\def\csname val@rev.desc.d1.uiv.p2.C.zoomclick-d3.delta-tau.tost1.equivalent\endcsname{0}
\expandafter\def\csname cnt@rev.desc.d1.uiv.p2.C.zoomclick-d3.delta-tau.tost1.equivalent\endcsname{0}
\expandafter\def\csname pct@rev.desc.d1.uiv.p2.C.zoomclick-d3.delta-tau.tost1.equivalent\endcsname{0.0}
\expandafter\def\csname ppt@rev.desc.d1.uiv.p2.C.zoomclick-d3.delta-tau.tost1.equivalent\endcsname{0.0}
\expandafter\def\csname val@rev.desc.d1.uiv.p2.C.zoomclick-d3.delta-tau.tost1.p\endcsname{0.645}
\expandafter\def\csname cnt@rev.desc.d1.uiv.p2.C.zoomclick-d3.delta-tau.tost1.p\endcsname{1}
\expandafter\def\csname pct@rev.desc.d1.uiv.p2.C.zoomclick-d3.delta-tau.tost1.p\endcsname{64.5}
\expandafter\def\csname ppt@rev.desc.d1.uiv.p2.C.zoomclick-d3.delta-tau.tost1.p\endcsname{$+$64.5}
\expandafter\def\csname val@rev.desc.d1.uiv.p2.C.zoomclick-d3.delta-tau.tost5.ci90-inside\endcsname{0}
\expandafter\def\csname cnt@rev.desc.d1.uiv.p2.C.zoomclick-d3.delta-tau.tost5.ci90-inside\endcsname{0}
\expandafter\def\csname pct@rev.desc.d1.uiv.p2.C.zoomclick-d3.delta-tau.tost5.ci90-inside\endcsname{0.0}
\expandafter\def\csname ppt@rev.desc.d1.uiv.p2.C.zoomclick-d3.delta-tau.tost5.ci90-inside\endcsname{0.0}
\expandafter\def\csname val@rev.desc.d1.uiv.p2.C.zoomclick-d3.delta-tau.tost5.equivalent\endcsname{0}
\expandafter\def\csname cnt@rev.desc.d1.uiv.p2.C.zoomclick-d3.delta-tau.tost5.equivalent\endcsname{0}
\expandafter\def\csname pct@rev.desc.d1.uiv.p2.C.zoomclick-d3.delta-tau.tost5.equivalent\endcsname{0.0}
\expandafter\def\csname ppt@rev.desc.d1.uiv.p2.C.zoomclick-d3.delta-tau.tost5.equivalent\endcsname{0.0}
\expandafter\def\csname val@rev.desc.d1.uiv.p2.C.zoomclick-d3.delta-tau.tost5.p\endcsname{0.384}
\expandafter\def\csname cnt@rev.desc.d1.uiv.p2.C.zoomclick-d3.delta-tau.tost5.p\endcsname{0}
\expandafter\def\csname pct@rev.desc.d1.uiv.p2.C.zoomclick-d3.delta-tau.tost5.p\endcsname{38.4}
\expandafter\def\csname ppt@rev.desc.d1.uiv.p2.C.zoomclick-d3.delta-tau.tost5.p\endcsname{$+$38.4}
\expandafter\def\csname val@rev.desc.d1.uiv.p2.C.zoomclick-d3.material-flip\endcsname{0}
\expandafter\def\csname cnt@rev.desc.d1.uiv.p2.C.zoomclick-d3.material-flip\endcsname{0}
\expandafter\def\csname pct@rev.desc.d1.uiv.p2.C.zoomclick-d3.material-flip\endcsname{0.0}
\expandafter\def\csname ppt@rev.desc.d1.uiv.p2.C.zoomclick-d3.material-flip\endcsname{0.0}
\expandafter\def\csname val@rev.desc.d1.uiv.p2.C.zoomclick-d3.n-apps\endcsname{12}
\expandafter\def\csname cnt@rev.desc.d1.uiv.p2.C.zoomclick-d3.n-apps\endcsname{12}
\expandafter\def\csname pct@rev.desc.d1.uiv.p2.C.zoomclick-d3.n-apps\endcsname{1200.0}
\expandafter\def\csname ppt@rev.desc.d1.uiv.p2.C.zoomclick-d3.n-apps\endcsname{$+$1200.0}
\expandafter\def\csname val@rev.desc.d1.uiv.p2.C.zoomclick-d3.n-tasks\endcsname{96}
\expandafter\def\csname cnt@rev.desc.d1.uiv.p2.C.zoomclick-d3.n-tasks\endcsname{96}
\expandafter\def\csname pct@rev.desc.d1.uiv.p2.C.zoomclick-d3.n-tasks\endcsname{9600.0}
\expandafter\def\csname ppt@rev.desc.d1.uiv.p2.C.zoomclick-d3.n-tasks\endcsname{$+$9600.0}
\expandafter\def\csname val@rev.desc.d1.uiv.p2.C.zoomclick-d3.sign-differs\endcsname{1}
\expandafter\def\csname cnt@rev.desc.d1.uiv.p2.C.zoomclick-d3.sign-differs\endcsname{1}
\expandafter\def\csname pct@rev.desc.d1.uiv.p2.C.zoomclick-d3.sign-differs\endcsname{100.0}
\expandafter\def\csname ppt@rev.desc.d1.uiv.p2.C.zoomclick-d3.sign-differs\endcsname{$+$100.0}
\expandafter\def\csname val@rev.desc.d1.uiv.p2.C.zoomclick-d3.tau-source.ci90.0\endcsname{-0.094}
\expandafter\def\csname cnt@rev.desc.d1.uiv.p2.C.zoomclick-d3.tau-source.ci90.0\endcsname{0}
\expandafter\def\csname pct@rev.desc.d1.uiv.p2.C.zoomclick-d3.tau-source.ci90.0\endcsname{$-$9.4}
\expandafter\def\csname ppt@rev.desc.d1.uiv.p2.C.zoomclick-d3.tau-source.ci90.0\endcsname{$-$9.4}
\expandafter\def\csname val@rev.desc.d1.uiv.p2.C.zoomclick-d3.tau-source.ci90.1\endcsname{0.042}
\expandafter\def\csname cnt@rev.desc.d1.uiv.p2.C.zoomclick-d3.tau-source.ci90.1\endcsname{0}
\expandafter\def\csname pct@rev.desc.d1.uiv.p2.C.zoomclick-d3.tau-source.ci90.1\endcsname{4.2}
\expandafter\def\csname ppt@rev.desc.d1.uiv.p2.C.zoomclick-d3.tau-source.ci90.1\endcsname{$+$4.2}
\expandafter\def\csname val@rev.desc.d1.uiv.p2.C.zoomclick-d3.tau-source.ci95-excludes-0\endcsname{0}
\expandafter\def\csname cnt@rev.desc.d1.uiv.p2.C.zoomclick-d3.tau-source.ci95-excludes-0\endcsname{0}
\expandafter\def\csname pct@rev.desc.d1.uiv.p2.C.zoomclick-d3.tau-source.ci95-excludes-0\endcsname{0.0}
\expandafter\def\csname ppt@rev.desc.d1.uiv.p2.C.zoomclick-d3.tau-source.ci95-excludes-0\endcsname{0.0}
\expandafter\def\csname val@rev.desc.d1.uiv.p2.C.zoomclick-d3.tau-source.ci95.0\endcsname{-0.104}
\expandafter\def\csname cnt@rev.desc.d1.uiv.p2.C.zoomclick-d3.tau-source.ci95.0\endcsname{0}
\expandafter\def\csname pct@rev.desc.d1.uiv.p2.C.zoomclick-d3.tau-source.ci95.0\endcsname{$-$10.4}
\expandafter\def\csname ppt@rev.desc.d1.uiv.p2.C.zoomclick-d3.tau-source.ci95.0\endcsname{$-$10.4}
\expandafter\def\csname val@rev.desc.d1.uiv.p2.C.zoomclick-d3.tau-source.ci95.1\endcsname{0.052}
\expandafter\def\csname cnt@rev.desc.d1.uiv.p2.C.zoomclick-d3.tau-source.ci95.1\endcsname{0}
\expandafter\def\csname pct@rev.desc.d1.uiv.p2.C.zoomclick-d3.tau-source.ci95.1\endcsname{5.2}
\expandafter\def\csname ppt@rev.desc.d1.uiv.p2.C.zoomclick-d3.tau-source.ci95.1\endcsname{$+$5.2}
\expandafter\def\csname val@rev.desc.d1.uiv.p2.C.zoomclick-d3.tau-source.mean\endcsname{-0.021}
\expandafter\def\csname cnt@rev.desc.d1.uiv.p2.C.zoomclick-d3.tau-source.mean\endcsname{0}
\expandafter\def\csname pct@rev.desc.d1.uiv.p2.C.zoomclick-d3.tau-source.mean\endcsname{$-$2.1}
\expandafter\def\csname ppt@rev.desc.d1.uiv.p2.C.zoomclick-d3.tau-source.mean\endcsname{$-$2.1}
\expandafter\def\csname val@rev.desc.d1.uiv.p2.C.zoomclick-d3.tau-source.n-apps\endcsname{12}
\expandafter\def\csname cnt@rev.desc.d1.uiv.p2.C.zoomclick-d3.tau-source.n-apps\endcsname{12}
\expandafter\def\csname pct@rev.desc.d1.uiv.p2.C.zoomclick-d3.tau-source.n-apps\endcsname{1200.0}
\expandafter\def\csname ppt@rev.desc.d1.uiv.p2.C.zoomclick-d3.tau-source.n-apps\endcsname{$+$1200.0}
\expandafter\def\csname val@rev.desc.d1.uiv.p2.C.zoomclick-d3.tau-source.tost1.ci90-inside\endcsname{0}
\expandafter\def\csname cnt@rev.desc.d1.uiv.p2.C.zoomclick-d3.tau-source.tost1.ci90-inside\endcsname{0}
\expandafter\def\csname pct@rev.desc.d1.uiv.p2.C.zoomclick-d3.tau-source.tost1.ci90-inside\endcsname{0.0}
\expandafter\def\csname ppt@rev.desc.d1.uiv.p2.C.zoomclick-d3.tau-source.tost1.ci90-inside\endcsname{0.0}
\expandafter\def\csname val@rev.desc.d1.uiv.p2.C.zoomclick-d3.tau-source.tost1.equivalent\endcsname{0}
\expandafter\def\csname cnt@rev.desc.d1.uiv.p2.C.zoomclick-d3.tau-source.tost1.equivalent\endcsname{0}
\expandafter\def\csname pct@rev.desc.d1.uiv.p2.C.zoomclick-d3.tau-source.tost1.equivalent\endcsname{0.0}
\expandafter\def\csname ppt@rev.desc.d1.uiv.p2.C.zoomclick-d3.tau-source.tost1.equivalent\endcsname{0.0}
\expandafter\def\csname val@rev.desc.d1.uiv.p2.C.zoomclick-d3.tau-source.tost1.p\endcsname{0.634}
\expandafter\def\csname cnt@rev.desc.d1.uiv.p2.C.zoomclick-d3.tau-source.tost1.p\endcsname{1}
\expandafter\def\csname pct@rev.desc.d1.uiv.p2.C.zoomclick-d3.tau-source.tost1.p\endcsname{63.4}
\expandafter\def\csname ppt@rev.desc.d1.uiv.p2.C.zoomclick-d3.tau-source.tost1.p\endcsname{$+$63.4}
\expandafter\def\csname val@rev.desc.d1.uiv.p2.C.zoomclick-d3.tau-source.tost5.ci90-inside\endcsname{0}
\expandafter\def\csname cnt@rev.desc.d1.uiv.p2.C.zoomclick-d3.tau-source.tost5.ci90-inside\endcsname{0}
\expandafter\def\csname pct@rev.desc.d1.uiv.p2.C.zoomclick-d3.tau-source.tost5.ci90-inside\endcsname{0.0}
\expandafter\def\csname ppt@rev.desc.d1.uiv.p2.C.zoomclick-d3.tau-source.tost5.ci90-inside\endcsname{0.0}
\expandafter\def\csname val@rev.desc.d1.uiv.p2.C.zoomclick-d3.tau-source.tost5.equivalent\endcsname{0}
\expandafter\def\csname cnt@rev.desc.d1.uiv.p2.C.zoomclick-d3.tau-source.tost5.equivalent\endcsname{0}
\expandafter\def\csname pct@rev.desc.d1.uiv.p2.C.zoomclick-d3.tau-source.tost5.equivalent\endcsname{0.0}
\expandafter\def\csname ppt@rev.desc.d1.uiv.p2.C.zoomclick-d3.tau-source.tost5.equivalent\endcsname{0.0}
\expandafter\def\csname val@rev.desc.d1.uiv.p2.C.zoomclick-d3.tau-source.tost5.p\endcsname{0.257}
\expandafter\def\csname cnt@rev.desc.d1.uiv.p2.C.zoomclick-d3.tau-source.tost5.p\endcsname{0}
\expandafter\def\csname pct@rev.desc.d1.uiv.p2.C.zoomclick-d3.tau-source.tost5.p\endcsname{25.7}
\expandafter\def\csname ppt@rev.desc.d1.uiv.p2.C.zoomclick-d3.tau-source.tost5.p\endcsname{$+$25.7}
\expandafter\def\csname val@rev.desc.d1.uiv.p2.C.zoomclick-d3.tau-target.ci90.0\endcsname{-0.042}
\expandafter\def\csname cnt@rev.desc.d1.uiv.p2.C.zoomclick-d3.tau-target.ci90.0\endcsname{0}
\expandafter\def\csname pct@rev.desc.d1.uiv.p2.C.zoomclick-d3.tau-target.ci90.0\endcsname{$-$4.2}
\expandafter\def\csname ppt@rev.desc.d1.uiv.p2.C.zoomclick-d3.tau-target.ci90.0\endcsname{$-$4.2}
\expandafter\def\csname val@rev.desc.d1.uiv.p2.C.zoomclick-d3.tau-target.ci90.1\endcsname{0.062}
\expandafter\def\csname cnt@rev.desc.d1.uiv.p2.C.zoomclick-d3.tau-target.ci90.1\endcsname{0}
\expandafter\def\csname pct@rev.desc.d1.uiv.p2.C.zoomclick-d3.tau-target.ci90.1\endcsname{6.2}
\expandafter\def\csname ppt@rev.desc.d1.uiv.p2.C.zoomclick-d3.tau-target.ci90.1\endcsname{$+$6.2}
\expandafter\def\csname val@rev.desc.d1.uiv.p2.C.zoomclick-d3.tau-target.ci95-excludes-0\endcsname{0}
\expandafter\def\csname cnt@rev.desc.d1.uiv.p2.C.zoomclick-d3.tau-target.ci95-excludes-0\endcsname{0}
\expandafter\def\csname pct@rev.desc.d1.uiv.p2.C.zoomclick-d3.tau-target.ci95-excludes-0\endcsname{0.0}
\expandafter\def\csname ppt@rev.desc.d1.uiv.p2.C.zoomclick-d3.tau-target.ci95-excludes-0\endcsname{0.0}
\expandafter\def\csname val@rev.desc.d1.uiv.p2.C.zoomclick-d3.tau-target.ci95.0\endcsname{-0.052}
\expandafter\def\csname cnt@rev.desc.d1.uiv.p2.C.zoomclick-d3.tau-target.ci95.0\endcsname{0}
\expandafter\def\csname pct@rev.desc.d1.uiv.p2.C.zoomclick-d3.tau-target.ci95.0\endcsname{$-$5.2}
\expandafter\def\csname ppt@rev.desc.d1.uiv.p2.C.zoomclick-d3.tau-target.ci95.0\endcsname{$-$5.2}
\expandafter\def\csname val@rev.desc.d1.uiv.p2.C.zoomclick-d3.tau-target.ci95.1\endcsname{0.073}
\expandafter\def\csname cnt@rev.desc.d1.uiv.p2.C.zoomclick-d3.tau-target.ci95.1\endcsname{0}
\expandafter\def\csname pct@rev.desc.d1.uiv.p2.C.zoomclick-d3.tau-target.ci95.1\endcsname{7.3}
\expandafter\def\csname ppt@rev.desc.d1.uiv.p2.C.zoomclick-d3.tau-target.ci95.1\endcsname{$+$7.3}
\expandafter\def\csname val@rev.desc.d1.uiv.p2.C.zoomclick-d3.tau-target.mean\endcsname{0.010}
\expandafter\def\csname cnt@rev.desc.d1.uiv.p2.C.zoomclick-d3.tau-target.mean\endcsname{0}
\expandafter\def\csname pct@rev.desc.d1.uiv.p2.C.zoomclick-d3.tau-target.mean\endcsname{1.0}
\expandafter\def\csname ppt@rev.desc.d1.uiv.p2.C.zoomclick-d3.tau-target.mean\endcsname{$+$1.0}
\expandafter\def\csname val@rev.desc.d1.uiv.p2.C.zoomclick-d3.tau-target.n-apps\endcsname{12}
\expandafter\def\csname cnt@rev.desc.d1.uiv.p2.C.zoomclick-d3.tau-target.n-apps\endcsname{12}
\expandafter\def\csname pct@rev.desc.d1.uiv.p2.C.zoomclick-d3.tau-target.n-apps\endcsname{1200.0}
\expandafter\def\csname ppt@rev.desc.d1.uiv.p2.C.zoomclick-d3.tau-target.n-apps\endcsname{$+$1200.0}
\expandafter\def\csname val@rev.desc.d1.uiv.p2.C.zoomclick-d3.tau-target.tost1.ci90-inside\endcsname{0}
\expandafter\def\csname cnt@rev.desc.d1.uiv.p2.C.zoomclick-d3.tau-target.tost1.ci90-inside\endcsname{0}
\expandafter\def\csname pct@rev.desc.d1.uiv.p2.C.zoomclick-d3.tau-target.tost1.ci90-inside\endcsname{0.0}
\expandafter\def\csname ppt@rev.desc.d1.uiv.p2.C.zoomclick-d3.tau-target.tost1.ci90-inside\endcsname{0.0}
\expandafter\def\csname val@rev.desc.d1.uiv.p2.C.zoomclick-d3.tau-target.tost1.equivalent\endcsname{0}
\expandafter\def\csname cnt@rev.desc.d1.uiv.p2.C.zoomclick-d3.tau-target.tost1.equivalent\endcsname{0}
\expandafter\def\csname pct@rev.desc.d1.uiv.p2.C.zoomclick-d3.tau-target.tost1.equivalent\endcsname{0.0}
\expandafter\def\csname ppt@rev.desc.d1.uiv.p2.C.zoomclick-d3.tau-target.tost1.equivalent\endcsname{0.0}
\expandafter\def\csname val@rev.desc.d1.uiv.p2.C.zoomclick-d3.tau-target.tost1.p\endcsname{0.583}
\expandafter\def\csname cnt@rev.desc.d1.uiv.p2.C.zoomclick-d3.tau-target.tost1.p\endcsname{1}
\expandafter\def\csname pct@rev.desc.d1.uiv.p2.C.zoomclick-d3.tau-target.tost1.p\endcsname{58.3}
\expandafter\def\csname ppt@rev.desc.d1.uiv.p2.C.zoomclick-d3.tau-target.tost1.p\endcsname{$+$58.3}
\expandafter\def\csname val@rev.desc.d1.uiv.p2.C.zoomclick-d3.tau-target.tost5.ci90-inside\endcsname{0}
\expandafter\def\csname cnt@rev.desc.d1.uiv.p2.C.zoomclick-d3.tau-target.tost5.ci90-inside\endcsname{0}
\expandafter\def\csname pct@rev.desc.d1.uiv.p2.C.zoomclick-d3.tau-target.tost5.ci90-inside\endcsname{0.0}
\expandafter\def\csname ppt@rev.desc.d1.uiv.p2.C.zoomclick-d3.tau-target.tost5.ci90-inside\endcsname{0.0}
\expandafter\def\csname val@rev.desc.d1.uiv.p2.C.zoomclick-d3.tau-target.tost5.equivalent\endcsname{0}
\expandafter\def\csname cnt@rev.desc.d1.uiv.p2.C.zoomclick-d3.tau-target.tost5.equivalent\endcsname{0}
\expandafter\def\csname pct@rev.desc.d1.uiv.p2.C.zoomclick-d3.tau-target.tost5.equivalent\endcsname{0.0}
\expandafter\def\csname ppt@rev.desc.d1.uiv.p2.C.zoomclick-d3.tau-target.tost5.equivalent\endcsname{0.0}
\expandafter\def\csname val@rev.desc.d1.uiv.p2.C.zoomclick-d3.tau-target.tost5.p\endcsname{0.143}
\expandafter\def\csname cnt@rev.desc.d1.uiv.p2.C.zoomclick-d3.tau-target.tost5.p\endcsname{0}
\expandafter\def\csname pct@rev.desc.d1.uiv.p2.C.zoomclick-d3.tau-target.tost5.p\endcsname{14.3}
\expandafter\def\csname ppt@rev.desc.d1.uiv.p2.C.zoomclick-d3.tau-target.tost5.p\endcsname{$+$14.3}
\expandafter\def\csname val@rev.desc.d1.uiv.p2.C.zoomclick-d3.zero-estimate\endcsname{0}
\expandafter\def\csname cnt@rev.desc.d1.uiv.p2.C.zoomclick-d3.zero-estimate\endcsname{0}
\expandafter\def\csname pct@rev.desc.d1.uiv.p2.C.zoomclick-d3.zero-estimate\endcsname{0.0}
\expandafter\def\csname ppt@rev.desc.d1.uiv.p2.C.zoomclick-d3.zero-estimate\endcsname{0.0}
\expandafter\def\csname val@rev.desc.d1.uiv.p2.T.center-d1.delta-tau.ci90.0\endcsname{-0.007}
\expandafter\def\csname cnt@rev.desc.d1.uiv.p2.T.center-d1.delta-tau.ci90.0\endcsname{0}
\expandafter\def\csname pct@rev.desc.d1.uiv.p2.T.center-d1.delta-tau.ci90.0\endcsname{$-$0.7}
\expandafter\def\csname ppt@rev.desc.d1.uiv.p2.T.center-d1.delta-tau.ci90.0\endcsname{$-$0.7}
\expandafter\def\csname val@rev.desc.d1.uiv.p2.T.center-d1.delta-tau.ci90.1\endcsname{0.062}
\expandafter\def\csname cnt@rev.desc.d1.uiv.p2.T.center-d1.delta-tau.ci90.1\endcsname{0}
\expandafter\def\csname pct@rev.desc.d1.uiv.p2.T.center-d1.delta-tau.ci90.1\endcsname{6.2}
\expandafter\def\csname ppt@rev.desc.d1.uiv.p2.T.center-d1.delta-tau.ci90.1\endcsname{$+$6.2}
\expandafter\def\csname val@rev.desc.d1.uiv.p2.T.center-d1.delta-tau.ci95-excludes-0\endcsname{0}
\expandafter\def\csname cnt@rev.desc.d1.uiv.p2.T.center-d1.delta-tau.ci95-excludes-0\endcsname{0}
\expandafter\def\csname pct@rev.desc.d1.uiv.p2.T.center-d1.delta-tau.ci95-excludes-0\endcsname{0.0}
\expandafter\def\csname ppt@rev.desc.d1.uiv.p2.T.center-d1.delta-tau.ci95-excludes-0\endcsname{0.0}
\expandafter\def\csname val@rev.desc.d1.uiv.p2.T.center-d1.delta-tau.ci95.0\endcsname{-0.014}
\expandafter\def\csname cnt@rev.desc.d1.uiv.p2.T.center-d1.delta-tau.ci95.0\endcsname{0}
\expandafter\def\csname pct@rev.desc.d1.uiv.p2.T.center-d1.delta-tau.ci95.0\endcsname{$-$1.4}
\expandafter\def\csname ppt@rev.desc.d1.uiv.p2.T.center-d1.delta-tau.ci95.0\endcsname{$-$1.4}
\expandafter\def\csname val@rev.desc.d1.uiv.p2.T.center-d1.delta-tau.ci95.1\endcsname{0.069}
\expandafter\def\csname cnt@rev.desc.d1.uiv.p2.T.center-d1.delta-tau.ci95.1\endcsname{0}
\expandafter\def\csname pct@rev.desc.d1.uiv.p2.T.center-d1.delta-tau.ci95.1\endcsname{6.9}
\expandafter\def\csname ppt@rev.desc.d1.uiv.p2.T.center-d1.delta-tau.ci95.1\endcsname{$+$6.9}
\expandafter\def\csname val@rev.desc.d1.uiv.p2.T.center-d1.delta-tau.mean\endcsname{0.028}
\expandafter\def\csname cnt@rev.desc.d1.uiv.p2.T.center-d1.delta-tau.mean\endcsname{0}
\expandafter\def\csname pct@rev.desc.d1.uiv.p2.T.center-d1.delta-tau.mean\endcsname{2.8}
\expandafter\def\csname ppt@rev.desc.d1.uiv.p2.T.center-d1.delta-tau.mean\endcsname{$+$2.8}
\expandafter\def\csname val@rev.desc.d1.uiv.p2.T.center-d1.delta-tau.n-apps\endcsname{24}
\expandafter\def\csname cnt@rev.desc.d1.uiv.p2.T.center-d1.delta-tau.n-apps\endcsname{24}
\expandafter\def\csname pct@rev.desc.d1.uiv.p2.T.center-d1.delta-tau.n-apps\endcsname{2400.0}
\expandafter\def\csname ppt@rev.desc.d1.uiv.p2.T.center-d1.delta-tau.n-apps\endcsname{$+$2400.0}
\expandafter\def\csname val@rev.desc.d1.uiv.p2.T.center-d1.delta-tau.tost1.ci90-inside\endcsname{0}
\expandafter\def\csname cnt@rev.desc.d1.uiv.p2.T.center-d1.delta-tau.tost1.ci90-inside\endcsname{0}
\expandafter\def\csname pct@rev.desc.d1.uiv.p2.T.center-d1.delta-tau.tost1.ci90-inside\endcsname{0.0}
\expandafter\def\csname ppt@rev.desc.d1.uiv.p2.T.center-d1.delta-tau.tost1.ci90-inside\endcsname{0.0}
\expandafter\def\csname val@rev.desc.d1.uiv.p2.T.center-d1.delta-tau.tost1.equivalent\endcsname{0}
\expandafter\def\csname cnt@rev.desc.d1.uiv.p2.T.center-d1.delta-tau.tost1.equivalent\endcsname{0}
\expandafter\def\csname pct@rev.desc.d1.uiv.p2.T.center-d1.delta-tau.tost1.equivalent\endcsname{0.0}
\expandafter\def\csname ppt@rev.desc.d1.uiv.p2.T.center-d1.delta-tau.tost1.equivalent\endcsname{0.0}
\expandafter\def\csname val@rev.desc.d1.uiv.p2.T.center-d1.delta-tau.tost1.p\endcsname{0.787}
\expandafter\def\csname cnt@rev.desc.d1.uiv.p2.T.center-d1.delta-tau.tost1.p\endcsname{1}
\expandafter\def\csname pct@rev.desc.d1.uiv.p2.T.center-d1.delta-tau.tost1.p\endcsname{78.7}
\expandafter\def\csname ppt@rev.desc.d1.uiv.p2.T.center-d1.delta-tau.tost1.p\endcsname{$+$78.7}
\expandafter\def\csname val@rev.desc.d1.uiv.p2.T.center-d1.delta-tau.tost5.ci90-inside\endcsname{0}
\expandafter\def\csname cnt@rev.desc.d1.uiv.p2.T.center-d1.delta-tau.tost5.ci90-inside\endcsname{0}
\expandafter\def\csname pct@rev.desc.d1.uiv.p2.T.center-d1.delta-tau.tost5.ci90-inside\endcsname{0.0}
\expandafter\def\csname ppt@rev.desc.d1.uiv.p2.T.center-d1.delta-tau.tost5.ci90-inside\endcsname{0.0}
\expandafter\def\csname val@rev.desc.d1.uiv.p2.T.center-d1.delta-tau.tost5.equivalent\endcsname{0}
\expandafter\def\csname cnt@rev.desc.d1.uiv.p2.T.center-d1.delta-tau.tost5.equivalent\endcsname{0}
\expandafter\def\csname pct@rev.desc.d1.uiv.p2.T.center-d1.delta-tau.tost5.equivalent\endcsname{0.0}
\expandafter\def\csname ppt@rev.desc.d1.uiv.p2.T.center-d1.delta-tau.tost5.equivalent\endcsname{0.0}
\expandafter\def\csname val@rev.desc.d1.uiv.p2.T.center-d1.delta-tau.tost5.p\endcsname{0.132}
\expandafter\def\csname cnt@rev.desc.d1.uiv.p2.T.center-d1.delta-tau.tost5.p\endcsname{0}
\expandafter\def\csname pct@rev.desc.d1.uiv.p2.T.center-d1.delta-tau.tost5.p\endcsname{13.2}
\expandafter\def\csname ppt@rev.desc.d1.uiv.p2.T.center-d1.delta-tau.tost5.p\endcsname{$+$13.2}
\expandafter\def\csname val@rev.desc.d1.uiv.p2.T.center-d1.material-flip\endcsname{0}
\expandafter\def\csname cnt@rev.desc.d1.uiv.p2.T.center-d1.material-flip\endcsname{0}
\expandafter\def\csname pct@rev.desc.d1.uiv.p2.T.center-d1.material-flip\endcsname{0.0}
\expandafter\def\csname ppt@rev.desc.d1.uiv.p2.T.center-d1.material-flip\endcsname{0.0}
\expandafter\def\csname val@rev.desc.d1.uiv.p2.T.center-d1.n-apps\endcsname{24}
\expandafter\def\csname cnt@rev.desc.d1.uiv.p2.T.center-d1.n-apps\endcsname{24}
\expandafter\def\csname pct@rev.desc.d1.uiv.p2.T.center-d1.n-apps\endcsname{2400.0}
\expandafter\def\csname ppt@rev.desc.d1.uiv.p2.T.center-d1.n-apps\endcsname{$+$2400.0}
\expandafter\def\csname val@rev.desc.d1.uiv.p2.T.center-d1.n-tasks\endcsname{144}
\expandafter\def\csname cnt@rev.desc.d1.uiv.p2.T.center-d1.n-tasks\endcsname{144}
\expandafter\def\csname pct@rev.desc.d1.uiv.p2.T.center-d1.n-tasks\endcsname{14400.0}
\expandafter\def\csname ppt@rev.desc.d1.uiv.p2.T.center-d1.n-tasks\endcsname{$+$14400.0}
\expandafter\def\csname val@rev.desc.d1.uiv.p2.T.center-d1.sign-differs\endcsname{0}
\expandafter\def\csname cnt@rev.desc.d1.uiv.p2.T.center-d1.sign-differs\endcsname{0}
\expandafter\def\csname pct@rev.desc.d1.uiv.p2.T.center-d1.sign-differs\endcsname{0.0}
\expandafter\def\csname ppt@rev.desc.d1.uiv.p2.T.center-d1.sign-differs\endcsname{0.0}
\expandafter\def\csname val@rev.desc.d1.uiv.p2.T.center-d1.tau-source.ci90.0\endcsname{-0.111}
\expandafter\def\csname cnt@rev.desc.d1.uiv.p2.T.center-d1.tau-source.ci90.0\endcsname{0}
\expandafter\def\csname pct@rev.desc.d1.uiv.p2.T.center-d1.tau-source.ci90.0\endcsname{$-$11.1}
\expandafter\def\csname ppt@rev.desc.d1.uiv.p2.T.center-d1.tau-source.ci90.0\endcsname{$-$11.1}
\expandafter\def\csname val@rev.desc.d1.uiv.p2.T.center-d1.tau-source.ci90.1\endcsname{-0.007}
\expandafter\def\csname cnt@rev.desc.d1.uiv.p2.T.center-d1.tau-source.ci90.1\endcsname{0}
\expandafter\def\csname pct@rev.desc.d1.uiv.p2.T.center-d1.tau-source.ci90.1\endcsname{$-$0.7}
\expandafter\def\csname ppt@rev.desc.d1.uiv.p2.T.center-d1.tau-source.ci90.1\endcsname{$-$0.7}
\expandafter\def\csname val@rev.desc.d1.uiv.p2.T.center-d1.tau-source.ci95-excludes-0\endcsname{1}
\expandafter\def\csname cnt@rev.desc.d1.uiv.p2.T.center-d1.tau-source.ci95-excludes-0\endcsname{1}
\expandafter\def\csname pct@rev.desc.d1.uiv.p2.T.center-d1.tau-source.ci95-excludes-0\endcsname{100.0}
\expandafter\def\csname ppt@rev.desc.d1.uiv.p2.T.center-d1.tau-source.ci95-excludes-0\endcsname{$+$100.0}
\expandafter\def\csname val@rev.desc.d1.uiv.p2.T.center-d1.tau-source.ci95.0\endcsname{-0.118}
\expandafter\def\csname cnt@rev.desc.d1.uiv.p2.T.center-d1.tau-source.ci95.0\endcsname{0}
\expandafter\def\csname pct@rev.desc.d1.uiv.p2.T.center-d1.tau-source.ci95.0\endcsname{$-$11.8}
\expandafter\def\csname ppt@rev.desc.d1.uiv.p2.T.center-d1.tau-source.ci95.0\endcsname{$-$11.8}
\expandafter\def\csname val@rev.desc.d1.uiv.p2.T.center-d1.tau-source.ci95.1\endcsname{-0.000}
\expandafter\def\csname cnt@rev.desc.d1.uiv.p2.T.center-d1.tau-source.ci95.1\endcsname{0}
\expandafter\def\csname pct@rev.desc.d1.uiv.p2.T.center-d1.tau-source.ci95.1\endcsname{0.0}
\expandafter\def\csname ppt@rev.desc.d1.uiv.p2.T.center-d1.tau-source.ci95.1\endcsname{0.0}
\expandafter\def\csname val@rev.desc.d1.uiv.p2.T.center-d1.tau-source.mean\endcsname{-0.056}
\expandafter\def\csname cnt@rev.desc.d1.uiv.p2.T.center-d1.tau-source.mean\endcsname{0}
\expandafter\def\csname pct@rev.desc.d1.uiv.p2.T.center-d1.tau-source.mean\endcsname{$-$5.6}
\expandafter\def\csname ppt@rev.desc.d1.uiv.p2.T.center-d1.tau-source.mean\endcsname{$-$5.6}
\expandafter\def\csname val@rev.desc.d1.uiv.p2.T.center-d1.tau-source.n-apps\endcsname{24}
\expandafter\def\csname cnt@rev.desc.d1.uiv.p2.T.center-d1.tau-source.n-apps\endcsname{24}
\expandafter\def\csname pct@rev.desc.d1.uiv.p2.T.center-d1.tau-source.n-apps\endcsname{2400.0}
\expandafter\def\csname ppt@rev.desc.d1.uiv.p2.T.center-d1.tau-source.n-apps\endcsname{$+$2400.0}
\expandafter\def\csname val@rev.desc.d1.uiv.p2.T.center-d1.tau-source.tost1.ci90-inside\endcsname{0}
\expandafter\def\csname cnt@rev.desc.d1.uiv.p2.T.center-d1.tau-source.tost1.ci90-inside\endcsname{0}
\expandafter\def\csname pct@rev.desc.d1.uiv.p2.T.center-d1.tau-source.tost1.ci90-inside\endcsname{0.0}
\expandafter\def\csname ppt@rev.desc.d1.uiv.p2.T.center-d1.tau-source.tost1.ci90-inside\endcsname{0.0}
\expandafter\def\csname val@rev.desc.d1.uiv.p2.T.center-d1.tau-source.tost1.equivalent\endcsname{0}
\expandafter\def\csname cnt@rev.desc.d1.uiv.p2.T.center-d1.tau-source.tost1.equivalent\endcsname{0}
\expandafter\def\csname pct@rev.desc.d1.uiv.p2.T.center-d1.tau-source.tost1.equivalent\endcsname{0.0}
\expandafter\def\csname ppt@rev.desc.d1.uiv.p2.T.center-d1.tau-source.tost1.equivalent\endcsname{0.0}
\expandafter\def\csname val@rev.desc.d1.uiv.p2.T.center-d1.tau-source.tost1.p\endcsname{0.939}
\expandafter\def\csname cnt@rev.desc.d1.uiv.p2.T.center-d1.tau-source.tost1.p\endcsname{1}
\expandafter\def\csname pct@rev.desc.d1.uiv.p2.T.center-d1.tau-source.tost1.p\endcsname{93.9}
\expandafter\def\csname ppt@rev.desc.d1.uiv.p2.T.center-d1.tau-source.tost1.p\endcsname{$+$93.9}
\expandafter\def\csname val@rev.desc.d1.uiv.p2.T.center-d1.tau-source.tost5.ci90-inside\endcsname{0}
\expandafter\def\csname cnt@rev.desc.d1.uiv.p2.T.center-d1.tau-source.tost5.ci90-inside\endcsname{0}
\expandafter\def\csname pct@rev.desc.d1.uiv.p2.T.center-d1.tau-source.tost5.ci90-inside\endcsname{0.0}
\expandafter\def\csname ppt@rev.desc.d1.uiv.p2.T.center-d1.tau-source.tost5.ci90-inside\endcsname{0.0}
\expandafter\def\csname val@rev.desc.d1.uiv.p2.T.center-d1.tau-source.tost5.equivalent\endcsname{0}
\expandafter\def\csname cnt@rev.desc.d1.uiv.p2.T.center-d1.tau-source.tost5.equivalent\endcsname{0}
\expandafter\def\csname pct@rev.desc.d1.uiv.p2.T.center-d1.tau-source.tost5.equivalent\endcsname{0.0}
\expandafter\def\csname ppt@rev.desc.d1.uiv.p2.T.center-d1.tau-source.tost5.equivalent\endcsname{0.0}
\expandafter\def\csname val@rev.desc.d1.uiv.p2.T.center-d1.tau-source.tost5.p\endcsname{0.528}
\expandafter\def\csname cnt@rev.desc.d1.uiv.p2.T.center-d1.tau-source.tost5.p\endcsname{1}
\expandafter\def\csname pct@rev.desc.d1.uiv.p2.T.center-d1.tau-source.tost5.p\endcsname{52.8}
\expandafter\def\csname ppt@rev.desc.d1.uiv.p2.T.center-d1.tau-source.tost5.p\endcsname{$+$52.8}
\expandafter\def\csname val@rev.desc.d1.uiv.p2.T.center-d1.tau-target.ci90.0\endcsname{-0.062}
\expandafter\def\csname cnt@rev.desc.d1.uiv.p2.T.center-d1.tau-target.ci90.0\endcsname{0}
\expandafter\def\csname pct@rev.desc.d1.uiv.p2.T.center-d1.tau-target.ci90.0\endcsname{$-$6.2}
\expandafter\def\csname ppt@rev.desc.d1.uiv.p2.T.center-d1.tau-target.ci90.0\endcsname{$-$6.2}
\expandafter\def\csname val@rev.desc.d1.uiv.p2.T.center-d1.tau-target.ci90.1\endcsname{0.007}
\expandafter\def\csname cnt@rev.desc.d1.uiv.p2.T.center-d1.tau-target.ci90.1\endcsname{0}
\expandafter\def\csname pct@rev.desc.d1.uiv.p2.T.center-d1.tau-target.ci90.1\endcsname{0.7}
\expandafter\def\csname ppt@rev.desc.d1.uiv.p2.T.center-d1.tau-target.ci90.1\endcsname{$+$0.7}
\expandafter\def\csname val@rev.desc.d1.uiv.p2.T.center-d1.tau-target.ci95-excludes-0\endcsname{0}
\expandafter\def\csname cnt@rev.desc.d1.uiv.p2.T.center-d1.tau-target.ci95-excludes-0\endcsname{0}
\expandafter\def\csname pct@rev.desc.d1.uiv.p2.T.center-d1.tau-target.ci95-excludes-0\endcsname{0.0}
\expandafter\def\csname ppt@rev.desc.d1.uiv.p2.T.center-d1.tau-target.ci95-excludes-0\endcsname{0.0}
\expandafter\def\csname val@rev.desc.d1.uiv.p2.T.center-d1.tau-target.ci95.0\endcsname{-0.069}
\expandafter\def\csname cnt@rev.desc.d1.uiv.p2.T.center-d1.tau-target.ci95.0\endcsname{0}
\expandafter\def\csname pct@rev.desc.d1.uiv.p2.T.center-d1.tau-target.ci95.0\endcsname{$-$6.9}
\expandafter\def\csname ppt@rev.desc.d1.uiv.p2.T.center-d1.tau-target.ci95.0\endcsname{$-$6.9}
\expandafter\def\csname val@rev.desc.d1.uiv.p2.T.center-d1.tau-target.ci95.1\endcsname{0.007}
\expandafter\def\csname cnt@rev.desc.d1.uiv.p2.T.center-d1.tau-target.ci95.1\endcsname{0}
\expandafter\def\csname pct@rev.desc.d1.uiv.p2.T.center-d1.tau-target.ci95.1\endcsname{0.7}
\expandafter\def\csname ppt@rev.desc.d1.uiv.p2.T.center-d1.tau-target.ci95.1\endcsname{$+$0.7}
\expandafter\def\csname val@rev.desc.d1.uiv.p2.T.center-d1.tau-target.mean\endcsname{-0.028}
\expandafter\def\csname cnt@rev.desc.d1.uiv.p2.T.center-d1.tau-target.mean\endcsname{0}
\expandafter\def\csname pct@rev.desc.d1.uiv.p2.T.center-d1.tau-target.mean\endcsname{$-$2.8}
\expandafter\def\csname ppt@rev.desc.d1.uiv.p2.T.center-d1.tau-target.mean\endcsname{$-$2.8}
\expandafter\def\csname val@rev.desc.d1.uiv.p2.T.center-d1.tau-target.n-apps\endcsname{24}
\expandafter\def\csname cnt@rev.desc.d1.uiv.p2.T.center-d1.tau-target.n-apps\endcsname{24}
\expandafter\def\csname pct@rev.desc.d1.uiv.p2.T.center-d1.tau-target.n-apps\endcsname{2400.0}
\expandafter\def\csname ppt@rev.desc.d1.uiv.p2.T.center-d1.tau-target.n-apps\endcsname{$+$2400.0}
\expandafter\def\csname val@rev.desc.d1.uiv.p2.T.center-d1.tau-target.tost1.ci90-inside\endcsname{0}
\expandafter\def\csname cnt@rev.desc.d1.uiv.p2.T.center-d1.tau-target.tost1.ci90-inside\endcsname{0}
\expandafter\def\csname pct@rev.desc.d1.uiv.p2.T.center-d1.tau-target.tost1.ci90-inside\endcsname{0.0}
\expandafter\def\csname ppt@rev.desc.d1.uiv.p2.T.center-d1.tau-target.tost1.ci90-inside\endcsname{0.0}
\expandafter\def\csname val@rev.desc.d1.uiv.p2.T.center-d1.tau-target.tost1.equivalent\endcsname{0}
\expandafter\def\csname cnt@rev.desc.d1.uiv.p2.T.center-d1.tau-target.tost1.equivalent\endcsname{0}
\expandafter\def\csname pct@rev.desc.d1.uiv.p2.T.center-d1.tau-target.tost1.equivalent\endcsname{0.0}
\expandafter\def\csname ppt@rev.desc.d1.uiv.p2.T.center-d1.tau-target.tost1.equivalent\endcsname{0.0}
\expandafter\def\csname val@rev.desc.d1.uiv.p2.T.center-d1.tau-target.tost1.p\endcsname{0.785}
\expandafter\def\csname cnt@rev.desc.d1.uiv.p2.T.center-d1.tau-target.tost1.p\endcsname{1}
\expandafter\def\csname pct@rev.desc.d1.uiv.p2.T.center-d1.tau-target.tost1.p\endcsname{78.5}
\expandafter\def\csname ppt@rev.desc.d1.uiv.p2.T.center-d1.tau-target.tost1.p\endcsname{$+$78.5}
\expandafter\def\csname val@rev.desc.d1.uiv.p2.T.center-d1.tau-target.tost5.ci90-inside\endcsname{0}
\expandafter\def\csname cnt@rev.desc.d1.uiv.p2.T.center-d1.tau-target.tost5.ci90-inside\endcsname{0}
\expandafter\def\csname pct@rev.desc.d1.uiv.p2.T.center-d1.tau-target.tost5.ci90-inside\endcsname{0.0}
\expandafter\def\csname ppt@rev.desc.d1.uiv.p2.T.center-d1.tau-target.tost5.ci90-inside\endcsname{0.0}
\expandafter\def\csname val@rev.desc.d1.uiv.p2.T.center-d1.tau-target.tost5.equivalent\endcsname{0}
\expandafter\def\csname cnt@rev.desc.d1.uiv.p2.T.center-d1.tau-target.tost5.equivalent\endcsname{0}
\expandafter\def\csname pct@rev.desc.d1.uiv.p2.T.center-d1.tau-target.tost5.equivalent\endcsname{0.0}
\expandafter\def\csname ppt@rev.desc.d1.uiv.p2.T.center-d1.tau-target.tost5.equivalent\endcsname{0.0}
\expandafter\def\csname val@rev.desc.d1.uiv.p2.T.center-d1.tau-target.tost5.p\endcsname{0.134}
\expandafter\def\csname cnt@rev.desc.d1.uiv.p2.T.center-d1.tau-target.tost5.p\endcsname{0}
\expandafter\def\csname pct@rev.desc.d1.uiv.p2.T.center-d1.tau-target.tost5.p\endcsname{13.4}
\expandafter\def\csname ppt@rev.desc.d1.uiv.p2.T.center-d1.tau-target.tost5.p\endcsname{$+$13.4}
\expandafter\def\csname val@rev.desc.d1.uiv.p2.T.center-d1.zero-estimate\endcsname{0}
\expandafter\def\csname cnt@rev.desc.d1.uiv.p2.T.center-d1.zero-estimate\endcsname{0}
\expandafter\def\csname pct@rev.desc.d1.uiv.p2.T.center-d1.zero-estimate\endcsname{0.0}
\expandafter\def\csname ppt@rev.desc.d1.uiv.p2.T.center-d1.zero-estimate\endcsname{0.0}
\expandafter\def\csname val@rev.desc.d1.uiv.p2.T.context-d1.delta-tau.ci90.0\endcsname{0.014}
\expandafter\def\csname cnt@rev.desc.d1.uiv.p2.T.context-d1.delta-tau.ci90.0\endcsname{0}
\expandafter\def\csname pct@rev.desc.d1.uiv.p2.T.context-d1.delta-tau.ci90.0\endcsname{1.4}
\expandafter\def\csname ppt@rev.desc.d1.uiv.p2.T.context-d1.delta-tau.ci90.0\endcsname{$+$1.4}
\expandafter\def\csname val@rev.desc.d1.uiv.p2.T.context-d1.delta-tau.ci90.1\endcsname{0.104}
\expandafter\def\csname cnt@rev.desc.d1.uiv.p2.T.context-d1.delta-tau.ci90.1\endcsname{0}
\expandafter\def\csname pct@rev.desc.d1.uiv.p2.T.context-d1.delta-tau.ci90.1\endcsname{10.4}
\expandafter\def\csname ppt@rev.desc.d1.uiv.p2.T.context-d1.delta-tau.ci90.1\endcsname{$+$10.4}
\expandafter\def\csname val@rev.desc.d1.uiv.p2.T.context-d1.delta-tau.ci95-excludes-0\endcsname{1}
\expandafter\def\csname cnt@rev.desc.d1.uiv.p2.T.context-d1.delta-tau.ci95-excludes-0\endcsname{1}
\expandafter\def\csname pct@rev.desc.d1.uiv.p2.T.context-d1.delta-tau.ci95-excludes-0\endcsname{100.0}
\expandafter\def\csname ppt@rev.desc.d1.uiv.p2.T.context-d1.delta-tau.ci95-excludes-0\endcsname{$+$100.0}
\expandafter\def\csname val@rev.desc.d1.uiv.p2.T.context-d1.delta-tau.ci95.0\endcsname{0.007}
\expandafter\def\csname cnt@rev.desc.d1.uiv.p2.T.context-d1.delta-tau.ci95.0\endcsname{0}
\expandafter\def\csname pct@rev.desc.d1.uiv.p2.T.context-d1.delta-tau.ci95.0\endcsname{0.7}
\expandafter\def\csname ppt@rev.desc.d1.uiv.p2.T.context-d1.delta-tau.ci95.0\endcsname{$+$0.7}
\expandafter\def\csname val@rev.desc.d1.uiv.p2.T.context-d1.delta-tau.ci95.1\endcsname{0.111}
\expandafter\def\csname cnt@rev.desc.d1.uiv.p2.T.context-d1.delta-tau.ci95.1\endcsname{0}
\expandafter\def\csname pct@rev.desc.d1.uiv.p2.T.context-d1.delta-tau.ci95.1\endcsname{11.1}
\expandafter\def\csname ppt@rev.desc.d1.uiv.p2.T.context-d1.delta-tau.ci95.1\endcsname{$+$11.1}
\expandafter\def\csname val@rev.desc.d1.uiv.p2.T.context-d1.delta-tau.mean\endcsname{0.056}
\expandafter\def\csname cnt@rev.desc.d1.uiv.p2.T.context-d1.delta-tau.mean\endcsname{0}
\expandafter\def\csname pct@rev.desc.d1.uiv.p2.T.context-d1.delta-tau.mean\endcsname{5.6}
\expandafter\def\csname ppt@rev.desc.d1.uiv.p2.T.context-d1.delta-tau.mean\endcsname{$+$5.6}
\expandafter\def\csname val@rev.desc.d1.uiv.p2.T.context-d1.delta-tau.n-apps\endcsname{24}
\expandafter\def\csname cnt@rev.desc.d1.uiv.p2.T.context-d1.delta-tau.n-apps\endcsname{24}
\expandafter\def\csname pct@rev.desc.d1.uiv.p2.T.context-d1.delta-tau.n-apps\endcsname{2400.0}
\expandafter\def\csname ppt@rev.desc.d1.uiv.p2.T.context-d1.delta-tau.n-apps\endcsname{$+$2400.0}
\expandafter\def\csname val@rev.desc.d1.uiv.p2.T.context-d1.delta-tau.tost1.ci90-inside\endcsname{0}
\expandafter\def\csname cnt@rev.desc.d1.uiv.p2.T.context-d1.delta-tau.tost1.ci90-inside\endcsname{0}
\expandafter\def\csname pct@rev.desc.d1.uiv.p2.T.context-d1.delta-tau.tost1.ci90-inside\endcsname{0.0}
\expandafter\def\csname ppt@rev.desc.d1.uiv.p2.T.context-d1.delta-tau.tost1.ci90-inside\endcsname{0.0}
\expandafter\def\csname val@rev.desc.d1.uiv.p2.T.context-d1.delta-tau.tost1.equivalent\endcsname{0}
\expandafter\def\csname cnt@rev.desc.d1.uiv.p2.T.context-d1.delta-tau.tost1.equivalent\endcsname{0}
\expandafter\def\csname pct@rev.desc.d1.uiv.p2.T.context-d1.delta-tau.tost1.equivalent\endcsname{0.0}
\expandafter\def\csname ppt@rev.desc.d1.uiv.p2.T.context-d1.delta-tau.tost1.equivalent\endcsname{0.0}
\expandafter\def\csname val@rev.desc.d1.uiv.p2.T.context-d1.delta-tau.tost1.p\endcsname{0.956}
\expandafter\def\csname cnt@rev.desc.d1.uiv.p2.T.context-d1.delta-tau.tost1.p\endcsname{1}
\expandafter\def\csname pct@rev.desc.d1.uiv.p2.T.context-d1.delta-tau.tost1.p\endcsname{95.6}
\expandafter\def\csname ppt@rev.desc.d1.uiv.p2.T.context-d1.delta-tau.tost1.p\endcsname{$+$95.6}
\expandafter\def\csname val@rev.desc.d1.uiv.p2.T.context-d1.delta-tau.tost5.ci90-inside\endcsname{0}
\expandafter\def\csname cnt@rev.desc.d1.uiv.p2.T.context-d1.delta-tau.tost5.ci90-inside\endcsname{0}
\expandafter\def\csname pct@rev.desc.d1.uiv.p2.T.context-d1.delta-tau.tost5.ci90-inside\endcsname{0.0}
\expandafter\def\csname ppt@rev.desc.d1.uiv.p2.T.context-d1.delta-tau.tost5.ci90-inside\endcsname{0.0}
\expandafter\def\csname val@rev.desc.d1.uiv.p2.T.context-d1.delta-tau.tost5.equivalent\endcsname{0}
\expandafter\def\csname cnt@rev.desc.d1.uiv.p2.T.context-d1.delta-tau.tost5.equivalent\endcsname{0}
\expandafter\def\csname pct@rev.desc.d1.uiv.p2.T.context-d1.delta-tau.tost5.equivalent\endcsname{0.0}
\expandafter\def\csname ppt@rev.desc.d1.uiv.p2.T.context-d1.delta-tau.tost5.equivalent\endcsname{0.0}
\expandafter\def\csname val@rev.desc.d1.uiv.p2.T.context-d1.delta-tau.tost5.p\endcsname{0.530}
\expandafter\def\csname cnt@rev.desc.d1.uiv.p2.T.context-d1.delta-tau.tost5.p\endcsname{1}
\expandafter\def\csname pct@rev.desc.d1.uiv.p2.T.context-d1.delta-tau.tost5.p\endcsname{53.0}
\expandafter\def\csname ppt@rev.desc.d1.uiv.p2.T.context-d1.delta-tau.tost5.p\endcsname{$+$53.0}
\expandafter\def\csname val@rev.desc.d1.uiv.p2.T.context-d1.material-flip\endcsname{0}
\expandafter\def\csname cnt@rev.desc.d1.uiv.p2.T.context-d1.material-flip\endcsname{0}
\expandafter\def\csname pct@rev.desc.d1.uiv.p2.T.context-d1.material-flip\endcsname{0.0}
\expandafter\def\csname ppt@rev.desc.d1.uiv.p2.T.context-d1.material-flip\endcsname{0.0}
\expandafter\def\csname val@rev.desc.d1.uiv.p2.T.context-d1.n-apps\endcsname{24}
\expandafter\def\csname cnt@rev.desc.d1.uiv.p2.T.context-d1.n-apps\endcsname{24}
\expandafter\def\csname pct@rev.desc.d1.uiv.p2.T.context-d1.n-apps\endcsname{2400.0}
\expandafter\def\csname ppt@rev.desc.d1.uiv.p2.T.context-d1.n-apps\endcsname{$+$2400.0}
\expandafter\def\csname val@rev.desc.d1.uiv.p2.T.context-d1.n-tasks\endcsname{144}
\expandafter\def\csname cnt@rev.desc.d1.uiv.p2.T.context-d1.n-tasks\endcsname{144}
\expandafter\def\csname pct@rev.desc.d1.uiv.p2.T.context-d1.n-tasks\endcsname{14400.0}
\expandafter\def\csname ppt@rev.desc.d1.uiv.p2.T.context-d1.n-tasks\endcsname{$+$14400.0}
\expandafter\def\csname val@rev.desc.d1.uiv.p2.T.context-d1.sign-differs\endcsname{0}
\expandafter\def\csname cnt@rev.desc.d1.uiv.p2.T.context-d1.sign-differs\endcsname{0}
\expandafter\def\csname pct@rev.desc.d1.uiv.p2.T.context-d1.sign-differs\endcsname{0.0}
\expandafter\def\csname ppt@rev.desc.d1.uiv.p2.T.context-d1.sign-differs\endcsname{0.0}
\expandafter\def\csname val@rev.desc.d1.uiv.p2.T.context-d1.tau-source.ci90.0\endcsname{-0.104}
\expandafter\def\csname cnt@rev.desc.d1.uiv.p2.T.context-d1.tau-source.ci90.0\endcsname{0}
\expandafter\def\csname pct@rev.desc.d1.uiv.p2.T.context-d1.tau-source.ci90.0\endcsname{$-$10.4}
\expandafter\def\csname ppt@rev.desc.d1.uiv.p2.T.context-d1.tau-source.ci90.0\endcsname{$-$10.4}
\expandafter\def\csname val@rev.desc.d1.uiv.p2.T.context-d1.tau-source.ci90.1\endcsname{-0.014}
\expandafter\def\csname cnt@rev.desc.d1.uiv.p2.T.context-d1.tau-source.ci90.1\endcsname{0}
\expandafter\def\csname pct@rev.desc.d1.uiv.p2.T.context-d1.tau-source.ci90.1\endcsname{$-$1.4}
\expandafter\def\csname ppt@rev.desc.d1.uiv.p2.T.context-d1.tau-source.ci90.1\endcsname{$-$1.4}
\expandafter\def\csname val@rev.desc.d1.uiv.p2.T.context-d1.tau-source.ci95-excludes-0\endcsname{1}
\expandafter\def\csname cnt@rev.desc.d1.uiv.p2.T.context-d1.tau-source.ci95-excludes-0\endcsname{1}
\expandafter\def\csname pct@rev.desc.d1.uiv.p2.T.context-d1.tau-source.ci95-excludes-0\endcsname{100.0}
\expandafter\def\csname ppt@rev.desc.d1.uiv.p2.T.context-d1.tau-source.ci95-excludes-0\endcsname{$+$100.0}
\expandafter\def\csname val@rev.desc.d1.uiv.p2.T.context-d1.tau-source.ci95.0\endcsname{-0.111}
\expandafter\def\csname cnt@rev.desc.d1.uiv.p2.T.context-d1.tau-source.ci95.0\endcsname{0}
\expandafter\def\csname pct@rev.desc.d1.uiv.p2.T.context-d1.tau-source.ci95.0\endcsname{$-$11.1}
\expandafter\def\csname ppt@rev.desc.d1.uiv.p2.T.context-d1.tau-source.ci95.0\endcsname{$-$11.1}
\expandafter\def\csname val@rev.desc.d1.uiv.p2.T.context-d1.tau-source.ci95.1\endcsname{-0.007}
\expandafter\def\csname cnt@rev.desc.d1.uiv.p2.T.context-d1.tau-source.ci95.1\endcsname{0}
\expandafter\def\csname pct@rev.desc.d1.uiv.p2.T.context-d1.tau-source.ci95.1\endcsname{$-$0.7}
\expandafter\def\csname ppt@rev.desc.d1.uiv.p2.T.context-d1.tau-source.ci95.1\endcsname{$-$0.7}
\expandafter\def\csname val@rev.desc.d1.uiv.p2.T.context-d1.tau-source.mean\endcsname{-0.056}
\expandafter\def\csname cnt@rev.desc.d1.uiv.p2.T.context-d1.tau-source.mean\endcsname{0}
\expandafter\def\csname pct@rev.desc.d1.uiv.p2.T.context-d1.tau-source.mean\endcsname{$-$5.6}
\expandafter\def\csname ppt@rev.desc.d1.uiv.p2.T.context-d1.tau-source.mean\endcsname{$-$5.6}
\expandafter\def\csname val@rev.desc.d1.uiv.p2.T.context-d1.tau-source.n-apps\endcsname{24}
\expandafter\def\csname cnt@rev.desc.d1.uiv.p2.T.context-d1.tau-source.n-apps\endcsname{24}
\expandafter\def\csname pct@rev.desc.d1.uiv.p2.T.context-d1.tau-source.n-apps\endcsname{2400.0}
\expandafter\def\csname ppt@rev.desc.d1.uiv.p2.T.context-d1.tau-source.n-apps\endcsname{$+$2400.0}
\expandafter\def\csname val@rev.desc.d1.uiv.p2.T.context-d1.tau-source.tost1.ci90-inside\endcsname{0}
\expandafter\def\csname cnt@rev.desc.d1.uiv.p2.T.context-d1.tau-source.tost1.ci90-inside\endcsname{0}
\expandafter\def\csname pct@rev.desc.d1.uiv.p2.T.context-d1.tau-source.tost1.ci90-inside\endcsname{0.0}
\expandafter\def\csname ppt@rev.desc.d1.uiv.p2.T.context-d1.tau-source.tost1.ci90-inside\endcsname{0.0}
\expandafter\def\csname val@rev.desc.d1.uiv.p2.T.context-d1.tau-source.tost1.equivalent\endcsname{0}
\expandafter\def\csname cnt@rev.desc.d1.uiv.p2.T.context-d1.tau-source.tost1.equivalent\endcsname{0}
\expandafter\def\csname pct@rev.desc.d1.uiv.p2.T.context-d1.tau-source.tost1.equivalent\endcsname{0.0}
\expandafter\def\csname ppt@rev.desc.d1.uiv.p2.T.context-d1.tau-source.tost1.equivalent\endcsname{0.0}
\expandafter\def\csname val@rev.desc.d1.uiv.p2.T.context-d1.tau-source.tost1.p\endcsname{0.968}
\expandafter\def\csname cnt@rev.desc.d1.uiv.p2.T.context-d1.tau-source.tost1.p\endcsname{1}
\expandafter\def\csname pct@rev.desc.d1.uiv.p2.T.context-d1.tau-source.tost1.p\endcsname{96.8}
\expandafter\def\csname ppt@rev.desc.d1.uiv.p2.T.context-d1.tau-source.tost1.p\endcsname{$+$96.8}
\expandafter\def\csname val@rev.desc.d1.uiv.p2.T.context-d1.tau-source.tost5.ci90-inside\endcsname{0}
\expandafter\def\csname cnt@rev.desc.d1.uiv.p2.T.context-d1.tau-source.tost5.ci90-inside\endcsname{0}
\expandafter\def\csname pct@rev.desc.d1.uiv.p2.T.context-d1.tau-source.tost5.ci90-inside\endcsname{0.0}
\expandafter\def\csname ppt@rev.desc.d1.uiv.p2.T.context-d1.tau-source.tost5.ci90-inside\endcsname{0.0}
\expandafter\def\csname val@rev.desc.d1.uiv.p2.T.context-d1.tau-source.tost5.equivalent\endcsname{0}
\expandafter\def\csname cnt@rev.desc.d1.uiv.p2.T.context-d1.tau-source.tost5.equivalent\endcsname{0}
\expandafter\def\csname pct@rev.desc.d1.uiv.p2.T.context-d1.tau-source.tost5.equivalent\endcsname{0.0}
\expandafter\def\csname ppt@rev.desc.d1.uiv.p2.T.context-d1.tau-source.tost5.equivalent\endcsname{0.0}
\expandafter\def\csname val@rev.desc.d1.uiv.p2.T.context-d1.tau-source.tost5.p\endcsname{0.516}
\expandafter\def\csname cnt@rev.desc.d1.uiv.p2.T.context-d1.tau-source.tost5.p\endcsname{1}
\expandafter\def\csname pct@rev.desc.d1.uiv.p2.T.context-d1.tau-source.tost5.p\endcsname{51.6}
\expandafter\def\csname ppt@rev.desc.d1.uiv.p2.T.context-d1.tau-source.tost5.p\endcsname{$+$51.6}
\expandafter\def\csname val@rev.desc.d1.uiv.p2.T.context-d1.tau-target.ci90.0\endcsname{-0.021}
\expandafter\def\csname cnt@rev.desc.d1.uiv.p2.T.context-d1.tau-target.ci90.0\endcsname{0}
\expandafter\def\csname pct@rev.desc.d1.uiv.p2.T.context-d1.tau-target.ci90.0\endcsname{$-$2.1}
\expandafter\def\csname ppt@rev.desc.d1.uiv.p2.T.context-d1.tau-target.ci90.0\endcsname{$-$2.1}
\expandafter\def\csname val@rev.desc.d1.uiv.p2.T.context-d1.tau-target.ci90.1\endcsname{0.021}
\expandafter\def\csname cnt@rev.desc.d1.uiv.p2.T.context-d1.tau-target.ci90.1\endcsname{0}
\expandafter\def\csname pct@rev.desc.d1.uiv.p2.T.context-d1.tau-target.ci90.1\endcsname{2.1}
\expandafter\def\csname ppt@rev.desc.d1.uiv.p2.T.context-d1.tau-target.ci90.1\endcsname{$+$2.1}
\expandafter\def\csname val@rev.desc.d1.uiv.p2.T.context-d1.tau-target.ci95-excludes-0\endcsname{0}
\expandafter\def\csname cnt@rev.desc.d1.uiv.p2.T.context-d1.tau-target.ci95-excludes-0\endcsname{0}
\expandafter\def\csname pct@rev.desc.d1.uiv.p2.T.context-d1.tau-target.ci95-excludes-0\endcsname{0.0}
\expandafter\def\csname ppt@rev.desc.d1.uiv.p2.T.context-d1.tau-target.ci95-excludes-0\endcsname{0.0}
\expandafter\def\csname val@rev.desc.d1.uiv.p2.T.context-d1.tau-target.ci95.0\endcsname{-0.028}
\expandafter\def\csname cnt@rev.desc.d1.uiv.p2.T.context-d1.tau-target.ci95.0\endcsname{0}
\expandafter\def\csname pct@rev.desc.d1.uiv.p2.T.context-d1.tau-target.ci95.0\endcsname{$-$2.8}
\expandafter\def\csname ppt@rev.desc.d1.uiv.p2.T.context-d1.tau-target.ci95.0\endcsname{$-$2.8}
\expandafter\def\csname val@rev.desc.d1.uiv.p2.T.context-d1.tau-target.ci95.1\endcsname{0.028}
\expandafter\def\csname cnt@rev.desc.d1.uiv.p2.T.context-d1.tau-target.ci95.1\endcsname{0}
\expandafter\def\csname pct@rev.desc.d1.uiv.p2.T.context-d1.tau-target.ci95.1\endcsname{2.8}
\expandafter\def\csname ppt@rev.desc.d1.uiv.p2.T.context-d1.tau-target.ci95.1\endcsname{$+$2.8}
\expandafter\def\csname val@rev.desc.d1.uiv.p2.T.context-d1.tau-target.mean\endcsname{0.000}
\expandafter\def\csname cnt@rev.desc.d1.uiv.p2.T.context-d1.tau-target.mean\endcsname{0}
\expandafter\def\csname pct@rev.desc.d1.uiv.p2.T.context-d1.tau-target.mean\endcsname{0.0}
\expandafter\def\csname ppt@rev.desc.d1.uiv.p2.T.context-d1.tau-target.mean\endcsname{0.0}
\expandafter\def\csname val@rev.desc.d1.uiv.p2.T.context-d1.tau-target.n-apps\endcsname{24}
\expandafter\def\csname cnt@rev.desc.d1.uiv.p2.T.context-d1.tau-target.n-apps\endcsname{24}
\expandafter\def\csname pct@rev.desc.d1.uiv.p2.T.context-d1.tau-target.n-apps\endcsname{2400.0}
\expandafter\def\csname ppt@rev.desc.d1.uiv.p2.T.context-d1.tau-target.n-apps\endcsname{$+$2400.0}
\expandafter\def\csname val@rev.desc.d1.uiv.p2.T.context-d1.tau-target.tost1.ci90-inside\endcsname{0}
\expandafter\def\csname cnt@rev.desc.d1.uiv.p2.T.context-d1.tau-target.tost1.ci90-inside\endcsname{0}
\expandafter\def\csname pct@rev.desc.d1.uiv.p2.T.context-d1.tau-target.tost1.ci90-inside\endcsname{0.0}
\expandafter\def\csname ppt@rev.desc.d1.uiv.p2.T.context-d1.tau-target.tost1.ci90-inside\endcsname{0.0}
\expandafter\def\csname val@rev.desc.d1.uiv.p2.T.context-d1.tau-target.tost1.equivalent\endcsname{0}
\expandafter\def\csname cnt@rev.desc.d1.uiv.p2.T.context-d1.tau-target.tost1.equivalent\endcsname{0}
\expandafter\def\csname pct@rev.desc.d1.uiv.p2.T.context-d1.tau-target.tost1.equivalent\endcsname{0.0}
\expandafter\def\csname ppt@rev.desc.d1.uiv.p2.T.context-d1.tau-target.tost1.equivalent\endcsname{0.0}
\expandafter\def\csname val@rev.desc.d1.uiv.p2.T.context-d1.tau-target.tost1.p\endcsname{0.216}
\expandafter\def\csname cnt@rev.desc.d1.uiv.p2.T.context-d1.tau-target.tost1.p\endcsname{0}
\expandafter\def\csname pct@rev.desc.d1.uiv.p2.T.context-d1.tau-target.tost1.p\endcsname{21.6}
\expandafter\def\csname ppt@rev.desc.d1.uiv.p2.T.context-d1.tau-target.tost1.p\endcsname{$+$21.6}
\expandafter\def\csname val@rev.desc.d1.uiv.p2.T.context-d1.tau-target.tost5.ci90-inside\endcsname{1}
\expandafter\def\csname cnt@rev.desc.d1.uiv.p2.T.context-d1.tau-target.tost5.ci90-inside\endcsname{1}
\expandafter\def\csname pct@rev.desc.d1.uiv.p2.T.context-d1.tau-target.tost5.ci90-inside\endcsname{100.0}
\expandafter\def\csname ppt@rev.desc.d1.uiv.p2.T.context-d1.tau-target.tost5.ci90-inside\endcsname{$+$100.0}
\expandafter\def\csname val@rev.desc.d1.uiv.p2.T.context-d1.tau-target.tost5.equivalent\endcsname{1}
\expandafter\def\csname cnt@rev.desc.d1.uiv.p2.T.context-d1.tau-target.tost5.equivalent\endcsname{1}
\expandafter\def\csname pct@rev.desc.d1.uiv.p2.T.context-d1.tau-target.tost5.equivalent\endcsname{100.0}
\expandafter\def\csname ppt@rev.desc.d1.uiv.p2.T.context-d1.tau-target.tost5.equivalent\endcsname{$+$100.0}
\expandafter\def\csname val@rev.desc.d1.uiv.p2.T.context-d1.tau-target.tost5.p\endcsname{0.000}
\expandafter\def\csname cnt@rev.desc.d1.uiv.p2.T.context-d1.tau-target.tost5.p\endcsname{0}
\expandafter\def\csname pct@rev.desc.d1.uiv.p2.T.context-d1.tau-target.tost5.p\endcsname{0.0}
\expandafter\def\csname ppt@rev.desc.d1.uiv.p2.T.context-d1.tau-target.tost5.p\endcsname{0.0}
\expandafter\def\csname val@rev.desc.d1.uiv.p2.T.context-d1.zero-estimate\endcsname{1}
\expandafter\def\csname cnt@rev.desc.d1.uiv.p2.T.context-d1.zero-estimate\endcsname{1}
\expandafter\def\csname pct@rev.desc.d1.uiv.p2.T.context-d1.zero-estimate\endcsname{100.0}
\expandafter\def\csname ppt@rev.desc.d1.uiv.p2.T.context-d1.zero-estimate\endcsname{$+$100.0}
\expandafter\def\csname val@rev.desc.d1.uiv.p2.T.dual-d2.delta-tau.ci90.0\endcsname{-0.007}
\expandafter\def\csname cnt@rev.desc.d1.uiv.p2.T.dual-d2.delta-tau.ci90.0\endcsname{0}
\expandafter\def\csname pct@rev.desc.d1.uiv.p2.T.dual-d2.delta-tau.ci90.0\endcsname{$-$0.7}
\expandafter\def\csname ppt@rev.desc.d1.uiv.p2.T.dual-d2.delta-tau.ci90.0\endcsname{$-$0.7}
\expandafter\def\csname val@rev.desc.d1.uiv.p2.T.dual-d2.delta-tau.ci90.1\endcsname{0.097}
\expandafter\def\csname cnt@rev.desc.d1.uiv.p2.T.dual-d2.delta-tau.ci90.1\endcsname{0}
\expandafter\def\csname pct@rev.desc.d1.uiv.p2.T.dual-d2.delta-tau.ci90.1\endcsname{9.7}
\expandafter\def\csname ppt@rev.desc.d1.uiv.p2.T.dual-d2.delta-tau.ci90.1\endcsname{$+$9.7}
\expandafter\def\csname val@rev.desc.d1.uiv.p2.T.dual-d2.delta-tau.ci95-excludes-0\endcsname{0}
\expandafter\def\csname cnt@rev.desc.d1.uiv.p2.T.dual-d2.delta-tau.ci95-excludes-0\endcsname{0}
\expandafter\def\csname pct@rev.desc.d1.uiv.p2.T.dual-d2.delta-tau.ci95-excludes-0\endcsname{0.0}
\expandafter\def\csname ppt@rev.desc.d1.uiv.p2.T.dual-d2.delta-tau.ci95-excludes-0\endcsname{0.0}
\expandafter\def\csname val@rev.desc.d1.uiv.p2.T.dual-d2.delta-tau.ci95.0\endcsname{-0.021}
\expandafter\def\csname cnt@rev.desc.d1.uiv.p2.T.dual-d2.delta-tau.ci95.0\endcsname{0}
\expandafter\def\csname pct@rev.desc.d1.uiv.p2.T.dual-d2.delta-tau.ci95.0\endcsname{$-$2.1}
\expandafter\def\csname ppt@rev.desc.d1.uiv.p2.T.dual-d2.delta-tau.ci95.0\endcsname{$-$2.1}
\expandafter\def\csname val@rev.desc.d1.uiv.p2.T.dual-d2.delta-tau.ci95.1\endcsname{0.111}
\expandafter\def\csname cnt@rev.desc.d1.uiv.p2.T.dual-d2.delta-tau.ci95.1\endcsname{0}
\expandafter\def\csname pct@rev.desc.d1.uiv.p2.T.dual-d2.delta-tau.ci95.1\endcsname{11.1}
\expandafter\def\csname ppt@rev.desc.d1.uiv.p2.T.dual-d2.delta-tau.ci95.1\endcsname{$+$11.1}
\expandafter\def\csname val@rev.desc.d1.uiv.p2.T.dual-d2.delta-tau.mean\endcsname{0.042}
\expandafter\def\csname cnt@rev.desc.d1.uiv.p2.T.dual-d2.delta-tau.mean\endcsname{0}
\expandafter\def\csname pct@rev.desc.d1.uiv.p2.T.dual-d2.delta-tau.mean\endcsname{4.2}
\expandafter\def\csname ppt@rev.desc.d1.uiv.p2.T.dual-d2.delta-tau.mean\endcsname{$+$4.2}
\expandafter\def\csname val@rev.desc.d1.uiv.p2.T.dual-d2.delta-tau.n-apps\endcsname{24}
\expandafter\def\csname cnt@rev.desc.d1.uiv.p2.T.dual-d2.delta-tau.n-apps\endcsname{24}
\expandafter\def\csname pct@rev.desc.d1.uiv.p2.T.dual-d2.delta-tau.n-apps\endcsname{2400.0}
\expandafter\def\csname ppt@rev.desc.d1.uiv.p2.T.dual-d2.delta-tau.n-apps\endcsname{$+$2400.0}
\expandafter\def\csname val@rev.desc.d1.uiv.p2.T.dual-d2.delta-tau.tost1.ci90-inside\endcsname{0}
\expandafter\def\csname cnt@rev.desc.d1.uiv.p2.T.dual-d2.delta-tau.tost1.ci90-inside\endcsname{0}
\expandafter\def\csname pct@rev.desc.d1.uiv.p2.T.dual-d2.delta-tau.tost1.ci90-inside\endcsname{0.0}
\expandafter\def\csname ppt@rev.desc.d1.uiv.p2.T.dual-d2.delta-tau.tost1.ci90-inside\endcsname{0.0}
\expandafter\def\csname val@rev.desc.d1.uiv.p2.T.dual-d2.delta-tau.tost1.equivalent\endcsname{0}
\expandafter\def\csname cnt@rev.desc.d1.uiv.p2.T.dual-d2.delta-tau.tost1.equivalent\endcsname{0}
\expandafter\def\csname pct@rev.desc.d1.uiv.p2.T.dual-d2.delta-tau.tost1.equivalent\endcsname{0.0}
\expandafter\def\csname ppt@rev.desc.d1.uiv.p2.T.dual-d2.delta-tau.tost1.equivalent\endcsname{0.0}
\expandafter\def\csname val@rev.desc.d1.uiv.p2.T.dual-d2.delta-tau.tost1.p\endcsname{0.837}
\expandafter\def\csname cnt@rev.desc.d1.uiv.p2.T.dual-d2.delta-tau.tost1.p\endcsname{1}
\expandafter\def\csname pct@rev.desc.d1.uiv.p2.T.dual-d2.delta-tau.tost1.p\endcsname{83.7}
\expandafter\def\csname ppt@rev.desc.d1.uiv.p2.T.dual-d2.delta-tau.tost1.p\endcsname{$+$83.7}
\expandafter\def\csname val@rev.desc.d1.uiv.p2.T.dual-d2.delta-tau.tost5.ci90-inside\endcsname{0}
\expandafter\def\csname cnt@rev.desc.d1.uiv.p2.T.dual-d2.delta-tau.tost5.ci90-inside\endcsname{0}
\expandafter\def\csname pct@rev.desc.d1.uiv.p2.T.dual-d2.delta-tau.tost5.ci90-inside\endcsname{0.0}
\expandafter\def\csname ppt@rev.desc.d1.uiv.p2.T.dual-d2.delta-tau.tost5.ci90-inside\endcsname{0.0}
\expandafter\def\csname val@rev.desc.d1.uiv.p2.T.dual-d2.delta-tau.tost5.equivalent\endcsname{0}
\expandafter\def\csname cnt@rev.desc.d1.uiv.p2.T.dual-d2.delta-tau.tost5.equivalent\endcsname{0}
\expandafter\def\csname pct@rev.desc.d1.uiv.p2.T.dual-d2.delta-tau.tost5.equivalent\endcsname{0.0}
\expandafter\def\csname ppt@rev.desc.d1.uiv.p2.T.dual-d2.delta-tau.tost5.equivalent\endcsname{0.0}
\expandafter\def\csname val@rev.desc.d1.uiv.p2.T.dual-d2.delta-tau.tost5.p\endcsname{0.369}
\expandafter\def\csname cnt@rev.desc.d1.uiv.p2.T.dual-d2.delta-tau.tost5.p\endcsname{0}
\expandafter\def\csname pct@rev.desc.d1.uiv.p2.T.dual-d2.delta-tau.tost5.p\endcsname{36.9}
\expandafter\def\csname ppt@rev.desc.d1.uiv.p2.T.dual-d2.delta-tau.tost5.p\endcsname{$+$36.9}
\expandafter\def\csname val@rev.desc.d1.uiv.p2.T.dual-d2.material-flip\endcsname{0}
\expandafter\def\csname cnt@rev.desc.d1.uiv.p2.T.dual-d2.material-flip\endcsname{0}
\expandafter\def\csname pct@rev.desc.d1.uiv.p2.T.dual-d2.material-flip\endcsname{0.0}
\expandafter\def\csname ppt@rev.desc.d1.uiv.p2.T.dual-d2.material-flip\endcsname{0.0}
\expandafter\def\csname val@rev.desc.d1.uiv.p2.T.dual-d2.n-apps\endcsname{24}
\expandafter\def\csname cnt@rev.desc.d1.uiv.p2.T.dual-d2.n-apps\endcsname{24}
\expandafter\def\csname pct@rev.desc.d1.uiv.p2.T.dual-d2.n-apps\endcsname{2400.0}
\expandafter\def\csname ppt@rev.desc.d1.uiv.p2.T.dual-d2.n-apps\endcsname{$+$2400.0}
\expandafter\def\csname val@rev.desc.d1.uiv.p2.T.dual-d2.n-tasks\endcsname{144}
\expandafter\def\csname cnt@rev.desc.d1.uiv.p2.T.dual-d2.n-tasks\endcsname{144}
\expandafter\def\csname pct@rev.desc.d1.uiv.p2.T.dual-d2.n-tasks\endcsname{14400.0}
\expandafter\def\csname ppt@rev.desc.d1.uiv.p2.T.dual-d2.n-tasks\endcsname{$+$14400.0}
\expandafter\def\csname val@rev.desc.d1.uiv.p2.T.dual-d2.sign-differs\endcsname{0}
\expandafter\def\csname cnt@rev.desc.d1.uiv.p2.T.dual-d2.sign-differs\endcsname{0}
\expandafter\def\csname pct@rev.desc.d1.uiv.p2.T.dual-d2.sign-differs\endcsname{0.0}
\expandafter\def\csname ppt@rev.desc.d1.uiv.p2.T.dual-d2.sign-differs\endcsname{0.0}
\expandafter\def\csname val@rev.desc.d1.uiv.p2.T.dual-d2.tau-source.ci90.0\endcsname{-0.146}
\expandafter\def\csname cnt@rev.desc.d1.uiv.p2.T.dual-d2.tau-source.ci90.0\endcsname{0}
\expandafter\def\csname pct@rev.desc.d1.uiv.p2.T.dual-d2.tau-source.ci90.0\endcsname{$-$14.6}
\expandafter\def\csname ppt@rev.desc.d1.uiv.p2.T.dual-d2.tau-source.ci90.0\endcsname{$-$14.6}
\expandafter\def\csname val@rev.desc.d1.uiv.p2.T.dual-d2.tau-source.ci90.1\endcsname{-0.062}
\expandafter\def\csname cnt@rev.desc.d1.uiv.p2.T.dual-d2.tau-source.ci90.1\endcsname{0}
\expandafter\def\csname pct@rev.desc.d1.uiv.p2.T.dual-d2.tau-source.ci90.1\endcsname{$-$6.2}
\expandafter\def\csname ppt@rev.desc.d1.uiv.p2.T.dual-d2.tau-source.ci90.1\endcsname{$-$6.2}
\expandafter\def\csname val@rev.desc.d1.uiv.p2.T.dual-d2.tau-source.ci95-excludes-0\endcsname{1}
\expandafter\def\csname cnt@rev.desc.d1.uiv.p2.T.dual-d2.tau-source.ci95-excludes-0\endcsname{1}
\expandafter\def\csname pct@rev.desc.d1.uiv.p2.T.dual-d2.tau-source.ci95-excludes-0\endcsname{100.0}
\expandafter\def\csname ppt@rev.desc.d1.uiv.p2.T.dual-d2.tau-source.ci95-excludes-0\endcsname{$+$100.0}
\expandafter\def\csname val@rev.desc.d1.uiv.p2.T.dual-d2.tau-source.ci95.0\endcsname{-0.160}
\expandafter\def\csname cnt@rev.desc.d1.uiv.p2.T.dual-d2.tau-source.ci95.0\endcsname{0}
\expandafter\def\csname pct@rev.desc.d1.uiv.p2.T.dual-d2.tau-source.ci95.0\endcsname{$-$16.0}
\expandafter\def\csname ppt@rev.desc.d1.uiv.p2.T.dual-d2.tau-source.ci95.0\endcsname{$-$16.0}
\expandafter\def\csname val@rev.desc.d1.uiv.p2.T.dual-d2.tau-source.ci95.1\endcsname{-0.049}
\expandafter\def\csname cnt@rev.desc.d1.uiv.p2.T.dual-d2.tau-source.ci95.1\endcsname{0}
\expandafter\def\csname pct@rev.desc.d1.uiv.p2.T.dual-d2.tau-source.ci95.1\endcsname{$-$4.9}
\expandafter\def\csname ppt@rev.desc.d1.uiv.p2.T.dual-d2.tau-source.ci95.1\endcsname{$-$4.9}
\expandafter\def\csname val@rev.desc.d1.uiv.p2.T.dual-d2.tau-source.mean\endcsname{-0.104}
\expandafter\def\csname cnt@rev.desc.d1.uiv.p2.T.dual-d2.tau-source.mean\endcsname{0}
\expandafter\def\csname pct@rev.desc.d1.uiv.p2.T.dual-d2.tau-source.mean\endcsname{$-$10.4}
\expandafter\def\csname ppt@rev.desc.d1.uiv.p2.T.dual-d2.tau-source.mean\endcsname{$-$10.4}
\expandafter\def\csname val@rev.desc.d1.uiv.p2.T.dual-d2.tau-source.n-apps\endcsname{24}
\expandafter\def\csname cnt@rev.desc.d1.uiv.p2.T.dual-d2.tau-source.n-apps\endcsname{24}
\expandafter\def\csname pct@rev.desc.d1.uiv.p2.T.dual-d2.tau-source.n-apps\endcsname{2400.0}
\expandafter\def\csname ppt@rev.desc.d1.uiv.p2.T.dual-d2.tau-source.n-apps\endcsname{$+$2400.0}
\expandafter\def\csname val@rev.desc.d1.uiv.p2.T.dual-d2.tau-source.tost1.ci90-inside\endcsname{0}
\expandafter\def\csname cnt@rev.desc.d1.uiv.p2.T.dual-d2.tau-source.tost1.ci90-inside\endcsname{0}
\expandafter\def\csname pct@rev.desc.d1.uiv.p2.T.dual-d2.tau-source.tost1.ci90-inside\endcsname{0.0}
\expandafter\def\csname ppt@rev.desc.d1.uiv.p2.T.dual-d2.tau-source.tost1.ci90-inside\endcsname{0.0}
\expandafter\def\csname val@rev.desc.d1.uiv.p2.T.dual-d2.tau-source.tost1.equivalent\endcsname{0}
\expandafter\def\csname cnt@rev.desc.d1.uiv.p2.T.dual-d2.tau-source.tost1.equivalent\endcsname{0}
\expandafter\def\csname pct@rev.desc.d1.uiv.p2.T.dual-d2.tau-source.tost1.equivalent\endcsname{0.0}
\expandafter\def\csname ppt@rev.desc.d1.uiv.p2.T.dual-d2.tau-source.tost1.equivalent\endcsname{0.0}
\expandafter\def\csname val@rev.desc.d1.uiv.p2.T.dual-d2.tau-source.tost1.p\endcsname{0.999}
\expandafter\def\csname cnt@rev.desc.d1.uiv.p2.T.dual-d2.tau-source.tost1.p\endcsname{1}
\expandafter\def\csname pct@rev.desc.d1.uiv.p2.T.dual-d2.tau-source.tost1.p\endcsname{99.9}
\expandafter\def\csname ppt@rev.desc.d1.uiv.p2.T.dual-d2.tau-source.tost1.p\endcsname{$+$99.9}
\expandafter\def\csname val@rev.desc.d1.uiv.p2.T.dual-d2.tau-source.tost5.ci90-inside\endcsname{0}
\expandafter\def\csname cnt@rev.desc.d1.uiv.p2.T.dual-d2.tau-source.tost5.ci90-inside\endcsname{0}
\expandafter\def\csname pct@rev.desc.d1.uiv.p2.T.dual-d2.tau-source.tost5.ci90-inside\endcsname{0.0}
\expandafter\def\csname ppt@rev.desc.d1.uiv.p2.T.dual-d2.tau-source.tost5.ci90-inside\endcsname{0.0}
\expandafter\def\csname val@rev.desc.d1.uiv.p2.T.dual-d2.tau-source.tost5.equivalent\endcsname{0}
\expandafter\def\csname cnt@rev.desc.d1.uiv.p2.T.dual-d2.tau-source.tost5.equivalent\endcsname{0}
\expandafter\def\csname pct@rev.desc.d1.uiv.p2.T.dual-d2.tau-source.tost5.equivalent\endcsname{0.0}
\expandafter\def\csname ppt@rev.desc.d1.uiv.p2.T.dual-d2.tau-source.tost5.equivalent\endcsname{0.0}
\expandafter\def\csname val@rev.desc.d1.uiv.p2.T.dual-d2.tau-source.tost5.p\endcsname{0.973}
\expandafter\def\csname cnt@rev.desc.d1.uiv.p2.T.dual-d2.tau-source.tost5.p\endcsname{1}
\expandafter\def\csname pct@rev.desc.d1.uiv.p2.T.dual-d2.tau-source.tost5.p\endcsname{97.3}
\expandafter\def\csname ppt@rev.desc.d1.uiv.p2.T.dual-d2.tau-source.tost5.p\endcsname{$+$97.3}
\expandafter\def\csname val@rev.desc.d1.uiv.p2.T.dual-d2.tau-target.ci90.0\endcsname{-0.111}
\expandafter\def\csname cnt@rev.desc.d1.uiv.p2.T.dual-d2.tau-target.ci90.0\endcsname{0}
\expandafter\def\csname pct@rev.desc.d1.uiv.p2.T.dual-d2.tau-target.ci90.0\endcsname{$-$11.1}
\expandafter\def\csname ppt@rev.desc.d1.uiv.p2.T.dual-d2.tau-target.ci90.0\endcsname{$-$11.1}
\expandafter\def\csname val@rev.desc.d1.uiv.p2.T.dual-d2.tau-target.ci90.1\endcsname{-0.014}
\expandafter\def\csname cnt@rev.desc.d1.uiv.p2.T.dual-d2.tau-target.ci90.1\endcsname{0}
\expandafter\def\csname pct@rev.desc.d1.uiv.p2.T.dual-d2.tau-target.ci90.1\endcsname{$-$1.4}
\expandafter\def\csname ppt@rev.desc.d1.uiv.p2.T.dual-d2.tau-target.ci90.1\endcsname{$-$1.4}
\expandafter\def\csname val@rev.desc.d1.uiv.p2.T.dual-d2.tau-target.ci95-excludes-0\endcsname{1}
\expandafter\def\csname cnt@rev.desc.d1.uiv.p2.T.dual-d2.tau-target.ci95-excludes-0\endcsname{1}
\expandafter\def\csname pct@rev.desc.d1.uiv.p2.T.dual-d2.tau-target.ci95-excludes-0\endcsname{100.0}
\expandafter\def\csname ppt@rev.desc.d1.uiv.p2.T.dual-d2.tau-target.ci95-excludes-0\endcsname{$+$100.0}
\expandafter\def\csname val@rev.desc.d1.uiv.p2.T.dual-d2.tau-target.ci95.0\endcsname{-0.125}
\expandafter\def\csname cnt@rev.desc.d1.uiv.p2.T.dual-d2.tau-target.ci95.0\endcsname{0}
\expandafter\def\csname pct@rev.desc.d1.uiv.p2.T.dual-d2.tau-target.ci95.0\endcsname{$-$12.5}
\expandafter\def\csname ppt@rev.desc.d1.uiv.p2.T.dual-d2.tau-target.ci95.0\endcsname{$-$12.5}
\expandafter\def\csname val@rev.desc.d1.uiv.p2.T.dual-d2.tau-target.ci95.1\endcsname{-0.007}
\expandafter\def\csname cnt@rev.desc.d1.uiv.p2.T.dual-d2.tau-target.ci95.1\endcsname{0}
\expandafter\def\csname pct@rev.desc.d1.uiv.p2.T.dual-d2.tau-target.ci95.1\endcsname{$-$0.7}
\expandafter\def\csname ppt@rev.desc.d1.uiv.p2.T.dual-d2.tau-target.ci95.1\endcsname{$-$0.7}
\expandafter\def\csname val@rev.desc.d1.uiv.p2.T.dual-d2.tau-target.mean\endcsname{-0.062}
\expandafter\def\csname cnt@rev.desc.d1.uiv.p2.T.dual-d2.tau-target.mean\endcsname{0}
\expandafter\def\csname pct@rev.desc.d1.uiv.p2.T.dual-d2.tau-target.mean\endcsname{$-$6.2}
\expandafter\def\csname ppt@rev.desc.d1.uiv.p2.T.dual-d2.tau-target.mean\endcsname{$-$6.2}
\expandafter\def\csname val@rev.desc.d1.uiv.p2.T.dual-d2.tau-target.n-apps\endcsname{24}
\expandafter\def\csname cnt@rev.desc.d1.uiv.p2.T.dual-d2.tau-target.n-apps\endcsname{24}
\expandafter\def\csname pct@rev.desc.d1.uiv.p2.T.dual-d2.tau-target.n-apps\endcsname{2400.0}
\expandafter\def\csname ppt@rev.desc.d1.uiv.p2.T.dual-d2.tau-target.n-apps\endcsname{$+$2400.0}
\expandafter\def\csname val@rev.desc.d1.uiv.p2.T.dual-d2.tau-target.tost1.ci90-inside\endcsname{0}
\expandafter\def\csname cnt@rev.desc.d1.uiv.p2.T.dual-d2.tau-target.tost1.ci90-inside\endcsname{0}
\expandafter\def\csname pct@rev.desc.d1.uiv.p2.T.dual-d2.tau-target.tost1.ci90-inside\endcsname{0.0}
\expandafter\def\csname ppt@rev.desc.d1.uiv.p2.T.dual-d2.tau-target.tost1.ci90-inside\endcsname{0.0}
\expandafter\def\csname val@rev.desc.d1.uiv.p2.T.dual-d2.tau-target.tost1.equivalent\endcsname{0}
\expandafter\def\csname cnt@rev.desc.d1.uiv.p2.T.dual-d2.tau-target.tost1.equivalent\endcsname{0}
\expandafter\def\csname pct@rev.desc.d1.uiv.p2.T.dual-d2.tau-target.tost1.equivalent\endcsname{0.0}
\expandafter\def\csname ppt@rev.desc.d1.uiv.p2.T.dual-d2.tau-target.tost1.equivalent\endcsname{0.0}
\expandafter\def\csname val@rev.desc.d1.uiv.p2.T.dual-d2.tau-target.tost1.p\endcsname{0.965}
\expandafter\def\csname cnt@rev.desc.d1.uiv.p2.T.dual-d2.tau-target.tost1.p\endcsname{1}
\expandafter\def\csname pct@rev.desc.d1.uiv.p2.T.dual-d2.tau-target.tost1.p\endcsname{96.5}
\expandafter\def\csname ppt@rev.desc.d1.uiv.p2.T.dual-d2.tau-target.tost1.p\endcsname{$+$96.5}
\expandafter\def\csname val@rev.desc.d1.uiv.p2.T.dual-d2.tau-target.tost5.ci90-inside\endcsname{0}
\expandafter\def\csname cnt@rev.desc.d1.uiv.p2.T.dual-d2.tau-target.tost5.ci90-inside\endcsname{0}
\expandafter\def\csname pct@rev.desc.d1.uiv.p2.T.dual-d2.tau-target.tost5.ci90-inside\endcsname{0.0}
\expandafter\def\csname ppt@rev.desc.d1.uiv.p2.T.dual-d2.tau-target.tost5.ci90-inside\endcsname{0.0}
\expandafter\def\csname val@rev.desc.d1.uiv.p2.T.dual-d2.tau-target.tost5.equivalent\endcsname{0}
\expandafter\def\csname cnt@rev.desc.d1.uiv.p2.T.dual-d2.tau-target.tost5.equivalent\endcsname{0}
\expandafter\def\csname pct@rev.desc.d1.uiv.p2.T.dual-d2.tau-target.tost5.equivalent\endcsname{0.0}
\expandafter\def\csname ppt@rev.desc.d1.uiv.p2.T.dual-d2.tau-target.tost5.equivalent\endcsname{0.0}
\expandafter\def\csname val@rev.desc.d1.uiv.p2.T.dual-d2.tau-target.tost5.p\endcsname{0.614}
\expandafter\def\csname cnt@rev.desc.d1.uiv.p2.T.dual-d2.tau-target.tost5.p\endcsname{1}
\expandafter\def\csname pct@rev.desc.d1.uiv.p2.T.dual-d2.tau-target.tost5.p\endcsname{61.4}
\expandafter\def\csname ppt@rev.desc.d1.uiv.p2.T.dual-d2.tau-target.tost5.p\endcsname{$+$61.4}
\expandafter\def\csname val@rev.desc.d1.uiv.p2.T.dual-d2.zero-estimate\endcsname{0}
\expandafter\def\csname cnt@rev.desc.d1.uiv.p2.T.dual-d2.zero-estimate\endcsname{0}
\expandafter\def\csname pct@rev.desc.d1.uiv.p2.T.dual-d2.zero-estimate\endcsname{0.0}
\expandafter\def\csname ppt@rev.desc.d1.uiv.p2.T.dual-d2.zero-estimate\endcsname{0.0}
\expandafter\def\csname val@rev.desc.d1.uiv.p2.T.grid2-d3.delta-tau.ci90.0\endcsname{0.014}
\expandafter\def\csname cnt@rev.desc.d1.uiv.p2.T.grid2-d3.delta-tau.ci90.0\endcsname{0}
\expandafter\def\csname pct@rev.desc.d1.uiv.p2.T.grid2-d3.delta-tau.ci90.0\endcsname{1.4}
\expandafter\def\csname ppt@rev.desc.d1.uiv.p2.T.grid2-d3.delta-tau.ci90.0\endcsname{$+$1.4}
\expandafter\def\csname val@rev.desc.d1.uiv.p2.T.grid2-d3.delta-tau.ci90.1\endcsname{0.132}
\expandafter\def\csname cnt@rev.desc.d1.uiv.p2.T.grid2-d3.delta-tau.ci90.1\endcsname{0}
\expandafter\def\csname pct@rev.desc.d1.uiv.p2.T.grid2-d3.delta-tau.ci90.1\endcsname{13.2}
\expandafter\def\csname ppt@rev.desc.d1.uiv.p2.T.grid2-d3.delta-tau.ci90.1\endcsname{$+$13.2}
\expandafter\def\csname val@rev.desc.d1.uiv.p2.T.grid2-d3.delta-tau.ci95-excludes-0\endcsname{1}
\expandafter\def\csname cnt@rev.desc.d1.uiv.p2.T.grid2-d3.delta-tau.ci95-excludes-0\endcsname{1}
\expandafter\def\csname pct@rev.desc.d1.uiv.p2.T.grid2-d3.delta-tau.ci95-excludes-0\endcsname{100.0}
\expandafter\def\csname ppt@rev.desc.d1.uiv.p2.T.grid2-d3.delta-tau.ci95-excludes-0\endcsname{$+$100.0}
\expandafter\def\csname val@rev.desc.d1.uiv.p2.T.grid2-d3.delta-tau.ci95.0\endcsname{0.007}
\expandafter\def\csname cnt@rev.desc.d1.uiv.p2.T.grid2-d3.delta-tau.ci95.0\endcsname{0}
\expandafter\def\csname pct@rev.desc.d1.uiv.p2.T.grid2-d3.delta-tau.ci95.0\endcsname{0.7}
\expandafter\def\csname ppt@rev.desc.d1.uiv.p2.T.grid2-d3.delta-tau.ci95.0\endcsname{$+$0.7}
\expandafter\def\csname val@rev.desc.d1.uiv.p2.T.grid2-d3.delta-tau.ci95.1\endcsname{0.139}
\expandafter\def\csname cnt@rev.desc.d1.uiv.p2.T.grid2-d3.delta-tau.ci95.1\endcsname{0}
\expandafter\def\csname pct@rev.desc.d1.uiv.p2.T.grid2-d3.delta-tau.ci95.1\endcsname{13.9}
\expandafter\def\csname ppt@rev.desc.d1.uiv.p2.T.grid2-d3.delta-tau.ci95.1\endcsname{$+$13.9}
\expandafter\def\csname val@rev.desc.d1.uiv.p2.T.grid2-d3.delta-tau.mean\endcsname{0.069}
\expandafter\def\csname cnt@rev.desc.d1.uiv.p2.T.grid2-d3.delta-tau.mean\endcsname{0}
\expandafter\def\csname pct@rev.desc.d1.uiv.p2.T.grid2-d3.delta-tau.mean\endcsname{6.9}
\expandafter\def\csname ppt@rev.desc.d1.uiv.p2.T.grid2-d3.delta-tau.mean\endcsname{$+$6.9}
\expandafter\def\csname val@rev.desc.d1.uiv.p2.T.grid2-d3.delta-tau.n-apps\endcsname{24}
\expandafter\def\csname cnt@rev.desc.d1.uiv.p2.T.grid2-d3.delta-tau.n-apps\endcsname{24}
\expandafter\def\csname pct@rev.desc.d1.uiv.p2.T.grid2-d3.delta-tau.n-apps\endcsname{2400.0}
\expandafter\def\csname ppt@rev.desc.d1.uiv.p2.T.grid2-d3.delta-tau.n-apps\endcsname{$+$2400.0}
\expandafter\def\csname val@rev.desc.d1.uiv.p2.T.grid2-d3.delta-tau.tost1.ci90-inside\endcsname{0}
\expandafter\def\csname cnt@rev.desc.d1.uiv.p2.T.grid2-d3.delta-tau.tost1.ci90-inside\endcsname{0}
\expandafter\def\csname pct@rev.desc.d1.uiv.p2.T.grid2-d3.delta-tau.tost1.ci90-inside\endcsname{0.0}
\expandafter\def\csname ppt@rev.desc.d1.uiv.p2.T.grid2-d3.delta-tau.tost1.ci90-inside\endcsname{0.0}
\expandafter\def\csname val@rev.desc.d1.uiv.p2.T.grid2-d3.delta-tau.tost1.equivalent\endcsname{0}
\expandafter\def\csname cnt@rev.desc.d1.uiv.p2.T.grid2-d3.delta-tau.tost1.equivalent\endcsname{0}
\expandafter\def\csname pct@rev.desc.d1.uiv.p2.T.grid2-d3.delta-tau.tost1.equivalent\endcsname{0.0}
\expandafter\def\csname ppt@rev.desc.d1.uiv.p2.T.grid2-d3.delta-tau.tost1.equivalent\endcsname{0.0}
\expandafter\def\csname val@rev.desc.d1.uiv.p2.T.grid2-d3.delta-tau.tost1.p\endcsname{0.966}
\expandafter\def\csname cnt@rev.desc.d1.uiv.p2.T.grid2-d3.delta-tau.tost1.p\endcsname{1}
\expandafter\def\csname pct@rev.desc.d1.uiv.p2.T.grid2-d3.delta-tau.tost1.p\endcsname{96.6}
\expandafter\def\csname ppt@rev.desc.d1.uiv.p2.T.grid2-d3.delta-tau.tost1.p\endcsname{$+$96.6}
\expandafter\def\csname val@rev.desc.d1.uiv.p2.T.grid2-d3.delta-tau.tost5.ci90-inside\endcsname{0}
\expandafter\def\csname cnt@rev.desc.d1.uiv.p2.T.grid2-d3.delta-tau.tost5.ci90-inside\endcsname{0}
\expandafter\def\csname pct@rev.desc.d1.uiv.p2.T.grid2-d3.delta-tau.tost5.ci90-inside\endcsname{0.0}
\expandafter\def\csname ppt@rev.desc.d1.uiv.p2.T.grid2-d3.delta-tau.tost5.ci90-inside\endcsname{0.0}
\expandafter\def\csname val@rev.desc.d1.uiv.p2.T.grid2-d3.delta-tau.tost5.equivalent\endcsname{0}
\expandafter\def\csname cnt@rev.desc.d1.uiv.p2.T.grid2-d3.delta-tau.tost5.equivalent\endcsname{0}
\expandafter\def\csname pct@rev.desc.d1.uiv.p2.T.grid2-d3.delta-tau.tost5.equivalent\endcsname{0.0}
\expandafter\def\csname ppt@rev.desc.d1.uiv.p2.T.grid2-d3.delta-tau.tost5.equivalent\endcsname{0.0}
\expandafter\def\csname val@rev.desc.d1.uiv.p2.T.grid2-d3.delta-tau.tost5.p\endcsname{0.689}
\expandafter\def\csname cnt@rev.desc.d1.uiv.p2.T.grid2-d3.delta-tau.tost5.p\endcsname{1}
\expandafter\def\csname pct@rev.desc.d1.uiv.p2.T.grid2-d3.delta-tau.tost5.p\endcsname{68.9}
\expandafter\def\csname ppt@rev.desc.d1.uiv.p2.T.grid2-d3.delta-tau.tost5.p\endcsname{$+$68.9}
\expandafter\def\csname val@rev.desc.d1.uiv.p2.T.grid2-d3.material-flip\endcsname{0}
\expandafter\def\csname cnt@rev.desc.d1.uiv.p2.T.grid2-d3.material-flip\endcsname{0}
\expandafter\def\csname pct@rev.desc.d1.uiv.p2.T.grid2-d3.material-flip\endcsname{0.0}
\expandafter\def\csname ppt@rev.desc.d1.uiv.p2.T.grid2-d3.material-flip\endcsname{0.0}
\expandafter\def\csname val@rev.desc.d1.uiv.p2.T.grid2-d3.n-apps\endcsname{24}
\expandafter\def\csname cnt@rev.desc.d1.uiv.p2.T.grid2-d3.n-apps\endcsname{24}
\expandafter\def\csname pct@rev.desc.d1.uiv.p2.T.grid2-d3.n-apps\endcsname{2400.0}
\expandafter\def\csname ppt@rev.desc.d1.uiv.p2.T.grid2-d3.n-apps\endcsname{$+$2400.0}
\expandafter\def\csname val@rev.desc.d1.uiv.p2.T.grid2-d3.n-tasks\endcsname{144}
\expandafter\def\csname cnt@rev.desc.d1.uiv.p2.T.grid2-d3.n-tasks\endcsname{144}
\expandafter\def\csname pct@rev.desc.d1.uiv.p2.T.grid2-d3.n-tasks\endcsname{14400.0}
\expandafter\def\csname ppt@rev.desc.d1.uiv.p2.T.grid2-d3.n-tasks\endcsname{$+$14400.0}
\expandafter\def\csname val@rev.desc.d1.uiv.p2.T.grid2-d3.sign-differs\endcsname{0}
\expandafter\def\csname cnt@rev.desc.d1.uiv.p2.T.grid2-d3.sign-differs\endcsname{0}
\expandafter\def\csname pct@rev.desc.d1.uiv.p2.T.grid2-d3.sign-differs\endcsname{0.0}
\expandafter\def\csname ppt@rev.desc.d1.uiv.p2.T.grid2-d3.sign-differs\endcsname{0.0}
\expandafter\def\csname val@rev.desc.d1.uiv.p2.T.grid2-d3.tau-source.ci90.0\endcsname{-0.125}
\expandafter\def\csname cnt@rev.desc.d1.uiv.p2.T.grid2-d3.tau-source.ci90.0\endcsname{0}
\expandafter\def\csname pct@rev.desc.d1.uiv.p2.T.grid2-d3.tau-source.ci90.0\endcsname{$-$12.5}
\expandafter\def\csname ppt@rev.desc.d1.uiv.p2.T.grid2-d3.tau-source.ci90.0\endcsname{$-$12.5}
\expandafter\def\csname val@rev.desc.d1.uiv.p2.T.grid2-d3.tau-source.ci90.1\endcsname{-0.028}
\expandafter\def\csname cnt@rev.desc.d1.uiv.p2.T.grid2-d3.tau-source.ci90.1\endcsname{0}
\expandafter\def\csname pct@rev.desc.d1.uiv.p2.T.grid2-d3.tau-source.ci90.1\endcsname{$-$2.8}
\expandafter\def\csname ppt@rev.desc.d1.uiv.p2.T.grid2-d3.tau-source.ci90.1\endcsname{$-$2.8}
\expandafter\def\csname val@rev.desc.d1.uiv.p2.T.grid2-d3.tau-source.ci95-excludes-0\endcsname{1}
\expandafter\def\csname cnt@rev.desc.d1.uiv.p2.T.grid2-d3.tau-source.ci95-excludes-0\endcsname{1}
\expandafter\def\csname pct@rev.desc.d1.uiv.p2.T.grid2-d3.tau-source.ci95-excludes-0\endcsname{100.0}
\expandafter\def\csname ppt@rev.desc.d1.uiv.p2.T.grid2-d3.tau-source.ci95-excludes-0\endcsname{$+$100.0}
\expandafter\def\csname val@rev.desc.d1.uiv.p2.T.grid2-d3.tau-source.ci95.0\endcsname{-0.139}
\expandafter\def\csname cnt@rev.desc.d1.uiv.p2.T.grid2-d3.tau-source.ci95.0\endcsname{0}
\expandafter\def\csname pct@rev.desc.d1.uiv.p2.T.grid2-d3.tau-source.ci95.0\endcsname{$-$13.9}
\expandafter\def\csname ppt@rev.desc.d1.uiv.p2.T.grid2-d3.tau-source.ci95.0\endcsname{$-$13.9}
\expandafter\def\csname val@rev.desc.d1.uiv.p2.T.grid2-d3.tau-source.ci95.1\endcsname{-0.021}
\expandafter\def\csname cnt@rev.desc.d1.uiv.p2.T.grid2-d3.tau-source.ci95.1\endcsname{0}
\expandafter\def\csname pct@rev.desc.d1.uiv.p2.T.grid2-d3.tau-source.ci95.1\endcsname{$-$2.1}
\expandafter\def\csname ppt@rev.desc.d1.uiv.p2.T.grid2-d3.tau-source.ci95.1\endcsname{$-$2.1}
\expandafter\def\csname val@rev.desc.d1.uiv.p2.T.grid2-d3.tau-source.mean\endcsname{-0.076}
\expandafter\def\csname cnt@rev.desc.d1.uiv.p2.T.grid2-d3.tau-source.mean\endcsname{0}
\expandafter\def\csname pct@rev.desc.d1.uiv.p2.T.grid2-d3.tau-source.mean\endcsname{$-$7.6}
\expandafter\def\csname ppt@rev.desc.d1.uiv.p2.T.grid2-d3.tau-source.mean\endcsname{$-$7.6}
\expandafter\def\csname val@rev.desc.d1.uiv.p2.T.grid2-d3.tau-source.n-apps\endcsname{24}
\expandafter\def\csname cnt@rev.desc.d1.uiv.p2.T.grid2-d3.tau-source.n-apps\endcsname{24}
\expandafter\def\csname pct@rev.desc.d1.uiv.p2.T.grid2-d3.tau-source.n-apps\endcsname{2400.0}
\expandafter\def\csname ppt@rev.desc.d1.uiv.p2.T.grid2-d3.tau-source.n-apps\endcsname{$+$2400.0}
\expandafter\def\csname val@rev.desc.d1.uiv.p2.T.grid2-d3.tau-source.tost1.ci90-inside\endcsname{0}
\expandafter\def\csname cnt@rev.desc.d1.uiv.p2.T.grid2-d3.tau-source.tost1.ci90-inside\endcsname{0}
\expandafter\def\csname pct@rev.desc.d1.uiv.p2.T.grid2-d3.tau-source.tost1.ci90-inside\endcsname{0.0}
\expandafter\def\csname ppt@rev.desc.d1.uiv.p2.T.grid2-d3.tau-source.tost1.ci90-inside\endcsname{0.0}
\expandafter\def\csname val@rev.desc.d1.uiv.p2.T.grid2-d3.tau-source.tost1.equivalent\endcsname{0}
\expandafter\def\csname cnt@rev.desc.d1.uiv.p2.T.grid2-d3.tau-source.tost1.equivalent\endcsname{0}
\expandafter\def\csname pct@rev.desc.d1.uiv.p2.T.grid2-d3.tau-source.tost1.equivalent\endcsname{0.0}
\expandafter\def\csname ppt@rev.desc.d1.uiv.p2.T.grid2-d3.tau-source.tost1.equivalent\endcsname{0.0}
\expandafter\def\csname val@rev.desc.d1.uiv.p2.T.grid2-d3.tau-source.tost1.p\endcsname{0.986}
\expandafter\def\csname cnt@rev.desc.d1.uiv.p2.T.grid2-d3.tau-source.tost1.p\endcsname{1}
\expandafter\def\csname pct@rev.desc.d1.uiv.p2.T.grid2-d3.tau-source.tost1.p\endcsname{98.6}
\expandafter\def\csname ppt@rev.desc.d1.uiv.p2.T.grid2-d3.tau-source.tost1.p\endcsname{$+$98.6}
\expandafter\def\csname val@rev.desc.d1.uiv.p2.T.grid2-d3.tau-source.tost5.ci90-inside\endcsname{0}
\expandafter\def\csname cnt@rev.desc.d1.uiv.p2.T.grid2-d3.tau-source.tost5.ci90-inside\endcsname{0}
\expandafter\def\csname pct@rev.desc.d1.uiv.p2.T.grid2-d3.tau-source.tost5.ci90-inside\endcsname{0.0}
\expandafter\def\csname ppt@rev.desc.d1.uiv.p2.T.grid2-d3.tau-source.tost5.ci90-inside\endcsname{0.0}
\expandafter\def\csname val@rev.desc.d1.uiv.p2.T.grid2-d3.tau-source.tost5.equivalent\endcsname{0}
\expandafter\def\csname cnt@rev.desc.d1.uiv.p2.T.grid2-d3.tau-source.tost5.equivalent\endcsname{0}
\expandafter\def\csname pct@rev.desc.d1.uiv.p2.T.grid2-d3.tau-source.tost5.equivalent\endcsname{0.0}
\expandafter\def\csname ppt@rev.desc.d1.uiv.p2.T.grid2-d3.tau-source.tost5.equivalent\endcsname{0.0}
\expandafter\def\csname val@rev.desc.d1.uiv.p2.T.grid2-d3.tau-source.tost5.p\endcsname{0.788}
\expandafter\def\csname cnt@rev.desc.d1.uiv.p2.T.grid2-d3.tau-source.tost5.p\endcsname{1}
\expandafter\def\csname pct@rev.desc.d1.uiv.p2.T.grid2-d3.tau-source.tost5.p\endcsname{78.8}
\expandafter\def\csname ppt@rev.desc.d1.uiv.p2.T.grid2-d3.tau-source.tost5.p\endcsname{$+$78.8}
\expandafter\def\csname val@rev.desc.d1.uiv.p2.T.grid2-d3.tau-target.ci90.0\endcsname{-0.035}
\expandafter\def\csname cnt@rev.desc.d1.uiv.p2.T.grid2-d3.tau-target.ci90.0\endcsname{0}
\expandafter\def\csname pct@rev.desc.d1.uiv.p2.T.grid2-d3.tau-target.ci90.0\endcsname{$-$3.5}
\expandafter\def\csname ppt@rev.desc.d1.uiv.p2.T.grid2-d3.tau-target.ci90.0\endcsname{$-$3.5}
\expandafter\def\csname val@rev.desc.d1.uiv.p2.T.grid2-d3.tau-target.ci90.1\endcsname{0.021}
\expandafter\def\csname cnt@rev.desc.d1.uiv.p2.T.grid2-d3.tau-target.ci90.1\endcsname{0}
\expandafter\def\csname pct@rev.desc.d1.uiv.p2.T.grid2-d3.tau-target.ci90.1\endcsname{2.1}
\expandafter\def\csname ppt@rev.desc.d1.uiv.p2.T.grid2-d3.tau-target.ci90.1\endcsname{$+$2.1}
\expandafter\def\csname val@rev.desc.d1.uiv.p2.T.grid2-d3.tau-target.ci95-excludes-0\endcsname{0}
\expandafter\def\csname cnt@rev.desc.d1.uiv.p2.T.grid2-d3.tau-target.ci95-excludes-0\endcsname{0}
\expandafter\def\csname pct@rev.desc.d1.uiv.p2.T.grid2-d3.tau-target.ci95-excludes-0\endcsname{0.0}
\expandafter\def\csname ppt@rev.desc.d1.uiv.p2.T.grid2-d3.tau-target.ci95-excludes-0\endcsname{0.0}
\expandafter\def\csname val@rev.desc.d1.uiv.p2.T.grid2-d3.tau-target.ci95.0\endcsname{-0.042}
\expandafter\def\csname cnt@rev.desc.d1.uiv.p2.T.grid2-d3.tau-target.ci95.0\endcsname{0}
\expandafter\def\csname pct@rev.desc.d1.uiv.p2.T.grid2-d3.tau-target.ci95.0\endcsname{$-$4.2}
\expandafter\def\csname ppt@rev.desc.d1.uiv.p2.T.grid2-d3.tau-target.ci95.0\endcsname{$-$4.2}
\expandafter\def\csname val@rev.desc.d1.uiv.p2.T.grid2-d3.tau-target.ci95.1\endcsname{0.028}
\expandafter\def\csname cnt@rev.desc.d1.uiv.p2.T.grid2-d3.tau-target.ci95.1\endcsname{0}
\expandafter\def\csname pct@rev.desc.d1.uiv.p2.T.grid2-d3.tau-target.ci95.1\endcsname{2.8}
\expandafter\def\csname ppt@rev.desc.d1.uiv.p2.T.grid2-d3.tau-target.ci95.1\endcsname{$+$2.8}
\expandafter\def\csname val@rev.desc.d1.uiv.p2.T.grid2-d3.tau-target.mean\endcsname{-0.007}
\expandafter\def\csname cnt@rev.desc.d1.uiv.p2.T.grid2-d3.tau-target.mean\endcsname{0}
\expandafter\def\csname pct@rev.desc.d1.uiv.p2.T.grid2-d3.tau-target.mean\endcsname{$-$0.7}
\expandafter\def\csname ppt@rev.desc.d1.uiv.p2.T.grid2-d3.tau-target.mean\endcsname{$-$0.7}
\expandafter\def\csname val@rev.desc.d1.uiv.p2.T.grid2-d3.tau-target.n-apps\endcsname{24}
\expandafter\def\csname cnt@rev.desc.d1.uiv.p2.T.grid2-d3.tau-target.n-apps\endcsname{24}
\expandafter\def\csname pct@rev.desc.d1.uiv.p2.T.grid2-d3.tau-target.n-apps\endcsname{2400.0}
\expandafter\def\csname ppt@rev.desc.d1.uiv.p2.T.grid2-d3.tau-target.n-apps\endcsname{$+$2400.0}
\expandafter\def\csname val@rev.desc.d1.uiv.p2.T.grid2-d3.tau-target.tost1.ci90-inside\endcsname{0}
\expandafter\def\csname cnt@rev.desc.d1.uiv.p2.T.grid2-d3.tau-target.tost1.ci90-inside\endcsname{0}
\expandafter\def\csname pct@rev.desc.d1.uiv.p2.T.grid2-d3.tau-target.tost1.ci90-inside\endcsname{0.0}
\expandafter\def\csname ppt@rev.desc.d1.uiv.p2.T.grid2-d3.tau-target.tost1.ci90-inside\endcsname{0.0}
\expandafter\def\csname val@rev.desc.d1.uiv.p2.T.grid2-d3.tau-target.tost1.equivalent\endcsname{0}
\expandafter\def\csname cnt@rev.desc.d1.uiv.p2.T.grid2-d3.tau-target.tost1.equivalent\endcsname{0}
\expandafter\def\csname pct@rev.desc.d1.uiv.p2.T.grid2-d3.tau-target.tost1.equivalent\endcsname{0.0}
\expandafter\def\csname ppt@rev.desc.d1.uiv.p2.T.grid2-d3.tau-target.tost1.equivalent\endcsname{0.0}
\expandafter\def\csname val@rev.desc.d1.uiv.p2.T.grid2-d3.tau-target.tost1.p\endcsname{0.430}
\expandafter\def\csname cnt@rev.desc.d1.uiv.p2.T.grid2-d3.tau-target.tost1.p\endcsname{0}
\expandafter\def\csname pct@rev.desc.d1.uiv.p2.T.grid2-d3.tau-target.tost1.p\endcsname{43.0}
\expandafter\def\csname ppt@rev.desc.d1.uiv.p2.T.grid2-d3.tau-target.tost1.p\endcsname{$+$43.0}
\expandafter\def\csname val@rev.desc.d1.uiv.p2.T.grid2-d3.tau-target.tost5.ci90-inside\endcsname{1}
\expandafter\def\csname cnt@rev.desc.d1.uiv.p2.T.grid2-d3.tau-target.tost5.ci90-inside\endcsname{1}
\expandafter\def\csname pct@rev.desc.d1.uiv.p2.T.grid2-d3.tau-target.tost5.ci90-inside\endcsname{100.0}
\expandafter\def\csname ppt@rev.desc.d1.uiv.p2.T.grid2-d3.tau-target.tost5.ci90-inside\endcsname{$+$100.0}
\expandafter\def\csname val@rev.desc.d1.uiv.p2.T.grid2-d3.tau-target.tost5.equivalent\endcsname{1}
\expandafter\def\csname cnt@rev.desc.d1.uiv.p2.T.grid2-d3.tau-target.tost5.equivalent\endcsname{1}
\expandafter\def\csname pct@rev.desc.d1.uiv.p2.T.grid2-d3.tau-target.tost5.equivalent\endcsname{100.0}
\expandafter\def\csname ppt@rev.desc.d1.uiv.p2.T.grid2-d3.tau-target.tost5.equivalent\endcsname{$+$100.0}
\expandafter\def\csname val@rev.desc.d1.uiv.p2.T.grid2-d3.tau-target.tost5.p\endcsname{0.006}
\expandafter\def\csname cnt@rev.desc.d1.uiv.p2.T.grid2-d3.tau-target.tost5.p\endcsname{0}
\expandafter\def\csname pct@rev.desc.d1.uiv.p2.T.grid2-d3.tau-target.tost5.p\endcsname{0.6}
\expandafter\def\csname ppt@rev.desc.d1.uiv.p2.T.grid2-d3.tau-target.tost5.p\endcsname{$+$0.6}
\expandafter\def\csname val@rev.desc.d1.uiv.p2.T.grid2-d3.zero-estimate\endcsname{0}
\expandafter\def\csname cnt@rev.desc.d1.uiv.p2.T.grid2-d3.zero-estimate\endcsname{0}
\expandafter\def\csname pct@rev.desc.d1.uiv.p2.T.grid2-d3.zero-estimate\endcsname{0.0}
\expandafter\def\csname ppt@rev.desc.d1.uiv.p2.T.grid2-d3.zero-estimate\endcsname{0.0}
\expandafter\def\csname val@rev.desc.d1.uiv.p2.T.zoomclick-d3.delta-tau.ci90.0\endcsname{0.014}
\expandafter\def\csname cnt@rev.desc.d1.uiv.p2.T.zoomclick-d3.delta-tau.ci90.0\endcsname{0}
\expandafter\def\csname pct@rev.desc.d1.uiv.p2.T.zoomclick-d3.delta-tau.ci90.0\endcsname{1.4}
\expandafter\def\csname ppt@rev.desc.d1.uiv.p2.T.zoomclick-d3.delta-tau.ci90.0\endcsname{$+$1.4}
\expandafter\def\csname val@rev.desc.d1.uiv.p2.T.zoomclick-d3.delta-tau.ci90.1\endcsname{0.083}
\expandafter\def\csname cnt@rev.desc.d1.uiv.p2.T.zoomclick-d3.delta-tau.ci90.1\endcsname{0}
\expandafter\def\csname pct@rev.desc.d1.uiv.p2.T.zoomclick-d3.delta-tau.ci90.1\endcsname{8.3}
\expandafter\def\csname ppt@rev.desc.d1.uiv.p2.T.zoomclick-d3.delta-tau.ci90.1\endcsname{$+$8.3}
\expandafter\def\csname val@rev.desc.d1.uiv.p2.T.zoomclick-d3.delta-tau.ci95-excludes-0\endcsname{1}
\expandafter\def\csname cnt@rev.desc.d1.uiv.p2.T.zoomclick-d3.delta-tau.ci95-excludes-0\endcsname{1}
\expandafter\def\csname pct@rev.desc.d1.uiv.p2.T.zoomclick-d3.delta-tau.ci95-excludes-0\endcsname{100.0}
\expandafter\def\csname ppt@rev.desc.d1.uiv.p2.T.zoomclick-d3.delta-tau.ci95-excludes-0\endcsname{$+$100.0}
\expandafter\def\csname val@rev.desc.d1.uiv.p2.T.zoomclick-d3.delta-tau.ci95.0\endcsname{0.007}
\expandafter\def\csname cnt@rev.desc.d1.uiv.p2.T.zoomclick-d3.delta-tau.ci95.0\endcsname{0}
\expandafter\def\csname pct@rev.desc.d1.uiv.p2.T.zoomclick-d3.delta-tau.ci95.0\endcsname{0.7}
\expandafter\def\csname ppt@rev.desc.d1.uiv.p2.T.zoomclick-d3.delta-tau.ci95.0\endcsname{$+$0.7}
\expandafter\def\csname val@rev.desc.d1.uiv.p2.T.zoomclick-d3.delta-tau.ci95.1\endcsname{0.097}
\expandafter\def\csname cnt@rev.desc.d1.uiv.p2.T.zoomclick-d3.delta-tau.ci95.1\endcsname{0}
\expandafter\def\csname pct@rev.desc.d1.uiv.p2.T.zoomclick-d3.delta-tau.ci95.1\endcsname{9.7}
\expandafter\def\csname ppt@rev.desc.d1.uiv.p2.T.zoomclick-d3.delta-tau.ci95.1\endcsname{$+$9.7}
\expandafter\def\csname val@rev.desc.d1.uiv.p2.T.zoomclick-d3.delta-tau.mean\endcsname{0.049}
\expandafter\def\csname cnt@rev.desc.d1.uiv.p2.T.zoomclick-d3.delta-tau.mean\endcsname{0}
\expandafter\def\csname pct@rev.desc.d1.uiv.p2.T.zoomclick-d3.delta-tau.mean\endcsname{4.9}
\expandafter\def\csname ppt@rev.desc.d1.uiv.p2.T.zoomclick-d3.delta-tau.mean\endcsname{$+$4.9}
\expandafter\def\csname val@rev.desc.d1.uiv.p2.T.zoomclick-d3.delta-tau.n-apps\endcsname{24}
\expandafter\def\csname cnt@rev.desc.d1.uiv.p2.T.zoomclick-d3.delta-tau.n-apps\endcsname{24}
\expandafter\def\csname pct@rev.desc.d1.uiv.p2.T.zoomclick-d3.delta-tau.n-apps\endcsname{2400.0}
\expandafter\def\csname ppt@rev.desc.d1.uiv.p2.T.zoomclick-d3.delta-tau.n-apps\endcsname{$+$2400.0}
\expandafter\def\csname val@rev.desc.d1.uiv.p2.T.zoomclick-d3.delta-tau.tost1.ci90-inside\endcsname{0}
\expandafter\def\csname cnt@rev.desc.d1.uiv.p2.T.zoomclick-d3.delta-tau.tost1.ci90-inside\endcsname{0}
\expandafter\def\csname pct@rev.desc.d1.uiv.p2.T.zoomclick-d3.delta-tau.tost1.ci90-inside\endcsname{0.0}
\expandafter\def\csname ppt@rev.desc.d1.uiv.p2.T.zoomclick-d3.delta-tau.tost1.ci90-inside\endcsname{0.0}
\expandafter\def\csname val@rev.desc.d1.uiv.p2.T.zoomclick-d3.delta-tau.tost1.equivalent\endcsname{0}
\expandafter\def\csname cnt@rev.desc.d1.uiv.p2.T.zoomclick-d3.delta-tau.tost1.equivalent\endcsname{0}
\expandafter\def\csname pct@rev.desc.d1.uiv.p2.T.zoomclick-d3.delta-tau.tost1.equivalent\endcsname{0.0}
\expandafter\def\csname ppt@rev.desc.d1.uiv.p2.T.zoomclick-d3.delta-tau.tost1.equivalent\endcsname{0.0}
\expandafter\def\csname val@rev.desc.d1.uiv.p2.T.zoomclick-d3.delta-tau.tost1.p\endcsname{0.954}
\expandafter\def\csname cnt@rev.desc.d1.uiv.p2.T.zoomclick-d3.delta-tau.tost1.p\endcsname{1}
\expandafter\def\csname pct@rev.desc.d1.uiv.p2.T.zoomclick-d3.delta-tau.tost1.p\endcsname{95.4}
\expandafter\def\csname ppt@rev.desc.d1.uiv.p2.T.zoomclick-d3.delta-tau.tost1.p\endcsname{$+$95.4}
\expandafter\def\csname val@rev.desc.d1.uiv.p2.T.zoomclick-d3.delta-tau.tost5.ci90-inside\endcsname{0}
\expandafter\def\csname cnt@rev.desc.d1.uiv.p2.T.zoomclick-d3.delta-tau.tost5.ci90-inside\endcsname{0}
\expandafter\def\csname pct@rev.desc.d1.uiv.p2.T.zoomclick-d3.delta-tau.tost5.ci90-inside\endcsname{0.0}
\expandafter\def\csname ppt@rev.desc.d1.uiv.p2.T.zoomclick-d3.delta-tau.tost5.ci90-inside\endcsname{0.0}
\expandafter\def\csname val@rev.desc.d1.uiv.p2.T.zoomclick-d3.delta-tau.tost5.equivalent\endcsname{0}
\expandafter\def\csname cnt@rev.desc.d1.uiv.p2.T.zoomclick-d3.delta-tau.tost5.equivalent\endcsname{0}
\expandafter\def\csname pct@rev.desc.d1.uiv.p2.T.zoomclick-d3.delta-tau.tost5.equivalent\endcsname{0.0}
\expandafter\def\csname ppt@rev.desc.d1.uiv.p2.T.zoomclick-d3.delta-tau.tost5.equivalent\endcsname{0.0}
\expandafter\def\csname val@rev.desc.d1.uiv.p2.T.zoomclick-d3.delta-tau.tost5.p\endcsname{0.428}
\expandafter\def\csname cnt@rev.desc.d1.uiv.p2.T.zoomclick-d3.delta-tau.tost5.p\endcsname{0}
\expandafter\def\csname pct@rev.desc.d1.uiv.p2.T.zoomclick-d3.delta-tau.tost5.p\endcsname{42.8}
\expandafter\def\csname ppt@rev.desc.d1.uiv.p2.T.zoomclick-d3.delta-tau.tost5.p\endcsname{$+$42.8}
\expandafter\def\csname val@rev.desc.d1.uiv.p2.T.zoomclick-d3.material-flip\endcsname{0}
\expandafter\def\csname cnt@rev.desc.d1.uiv.p2.T.zoomclick-d3.material-flip\endcsname{0}
\expandafter\def\csname pct@rev.desc.d1.uiv.p2.T.zoomclick-d3.material-flip\endcsname{0.0}
\expandafter\def\csname ppt@rev.desc.d1.uiv.p2.T.zoomclick-d3.material-flip\endcsname{0.0}
\expandafter\def\csname val@rev.desc.d1.uiv.p2.T.zoomclick-d3.n-apps\endcsname{24}
\expandafter\def\csname cnt@rev.desc.d1.uiv.p2.T.zoomclick-d3.n-apps\endcsname{24}
\expandafter\def\csname pct@rev.desc.d1.uiv.p2.T.zoomclick-d3.n-apps\endcsname{2400.0}
\expandafter\def\csname ppt@rev.desc.d1.uiv.p2.T.zoomclick-d3.n-apps\endcsname{$+$2400.0}
\expandafter\def\csname val@rev.desc.d1.uiv.p2.T.zoomclick-d3.n-tasks\endcsname{144}
\expandafter\def\csname cnt@rev.desc.d1.uiv.p2.T.zoomclick-d3.n-tasks\endcsname{144}
\expandafter\def\csname pct@rev.desc.d1.uiv.p2.T.zoomclick-d3.n-tasks\endcsname{14400.0}
\expandafter\def\csname ppt@rev.desc.d1.uiv.p2.T.zoomclick-d3.n-tasks\endcsname{$+$14400.0}
\expandafter\def\csname val@rev.desc.d1.uiv.p2.T.zoomclick-d3.sign-differs\endcsname{0}
\expandafter\def\csname cnt@rev.desc.d1.uiv.p2.T.zoomclick-d3.sign-differs\endcsname{0}
\expandafter\def\csname pct@rev.desc.d1.uiv.p2.T.zoomclick-d3.sign-differs\endcsname{0.0}
\expandafter\def\csname ppt@rev.desc.d1.uiv.p2.T.zoomclick-d3.sign-differs\endcsname{0.0}
\expandafter\def\csname val@rev.desc.d1.uiv.p2.T.zoomclick-d3.tau-source.ci90.0\endcsname{-0.104}
\expandafter\def\csname cnt@rev.desc.d1.uiv.p2.T.zoomclick-d3.tau-source.ci90.0\endcsname{0}
\expandafter\def\csname pct@rev.desc.d1.uiv.p2.T.zoomclick-d3.tau-source.ci90.0\endcsname{$-$10.4}
\expandafter\def\csname ppt@rev.desc.d1.uiv.p2.T.zoomclick-d3.tau-source.ci90.0\endcsname{$-$10.4}
\expandafter\def\csname val@rev.desc.d1.uiv.p2.T.zoomclick-d3.tau-source.ci90.1\endcsname{-0.014}
\expandafter\def\csname cnt@rev.desc.d1.uiv.p2.T.zoomclick-d3.tau-source.ci90.1\endcsname{0}
\expandafter\def\csname pct@rev.desc.d1.uiv.p2.T.zoomclick-d3.tau-source.ci90.1\endcsname{$-$1.4}
\expandafter\def\csname ppt@rev.desc.d1.uiv.p2.T.zoomclick-d3.tau-source.ci90.1\endcsname{$-$1.4}
\expandafter\def\csname val@rev.desc.d1.uiv.p2.T.zoomclick-d3.tau-source.ci95-excludes-0\endcsname{1}
\expandafter\def\csname cnt@rev.desc.d1.uiv.p2.T.zoomclick-d3.tau-source.ci95-excludes-0\endcsname{1}
\expandafter\def\csname pct@rev.desc.d1.uiv.p2.T.zoomclick-d3.tau-source.ci95-excludes-0\endcsname{100.0}
\expandafter\def\csname ppt@rev.desc.d1.uiv.p2.T.zoomclick-d3.tau-source.ci95-excludes-0\endcsname{$+$100.0}
\expandafter\def\csname val@rev.desc.d1.uiv.p2.T.zoomclick-d3.tau-source.ci95.0\endcsname{-0.111}
\expandafter\def\csname cnt@rev.desc.d1.uiv.p2.T.zoomclick-d3.tau-source.ci95.0\endcsname{0}
\expandafter\def\csname pct@rev.desc.d1.uiv.p2.T.zoomclick-d3.tau-source.ci95.0\endcsname{$-$11.1}
\expandafter\def\csname ppt@rev.desc.d1.uiv.p2.T.zoomclick-d3.tau-source.ci95.0\endcsname{$-$11.1}
\expandafter\def\csname val@rev.desc.d1.uiv.p2.T.zoomclick-d3.tau-source.ci95.1\endcsname{-0.007}
\expandafter\def\csname cnt@rev.desc.d1.uiv.p2.T.zoomclick-d3.tau-source.ci95.1\endcsname{0}
\expandafter\def\csname pct@rev.desc.d1.uiv.p2.T.zoomclick-d3.tau-source.ci95.1\endcsname{$-$0.7}
\expandafter\def\csname ppt@rev.desc.d1.uiv.p2.T.zoomclick-d3.tau-source.ci95.1\endcsname{$-$0.7}
\expandafter\def\csname val@rev.desc.d1.uiv.p2.T.zoomclick-d3.tau-source.mean\endcsname{-0.056}
\expandafter\def\csname cnt@rev.desc.d1.uiv.p2.T.zoomclick-d3.tau-source.mean\endcsname{0}
\expandafter\def\csname pct@rev.desc.d1.uiv.p2.T.zoomclick-d3.tau-source.mean\endcsname{$-$5.6}
\expandafter\def\csname ppt@rev.desc.d1.uiv.p2.T.zoomclick-d3.tau-source.mean\endcsname{$-$5.6}
\expandafter\def\csname val@rev.desc.d1.uiv.p2.T.zoomclick-d3.tau-source.n-apps\endcsname{24}
\expandafter\def\csname cnt@rev.desc.d1.uiv.p2.T.zoomclick-d3.tau-source.n-apps\endcsname{24}
\expandafter\def\csname pct@rev.desc.d1.uiv.p2.T.zoomclick-d3.tau-source.n-apps\endcsname{2400.0}
\expandafter\def\csname ppt@rev.desc.d1.uiv.p2.T.zoomclick-d3.tau-source.n-apps\endcsname{$+$2400.0}
\expandafter\def\csname val@rev.desc.d1.uiv.p2.T.zoomclick-d3.tau-source.tost1.ci90-inside\endcsname{0}
\expandafter\def\csname cnt@rev.desc.d1.uiv.p2.T.zoomclick-d3.tau-source.tost1.ci90-inside\endcsname{0}
\expandafter\def\csname pct@rev.desc.d1.uiv.p2.T.zoomclick-d3.tau-source.tost1.ci90-inside\endcsname{0.0}
\expandafter\def\csname ppt@rev.desc.d1.uiv.p2.T.zoomclick-d3.tau-source.tost1.ci90-inside\endcsname{0.0}
\expandafter\def\csname val@rev.desc.d1.uiv.p2.T.zoomclick-d3.tau-source.tost1.equivalent\endcsname{0}
\expandafter\def\csname cnt@rev.desc.d1.uiv.p2.T.zoomclick-d3.tau-source.tost1.equivalent\endcsname{0}
\expandafter\def\csname pct@rev.desc.d1.uiv.p2.T.zoomclick-d3.tau-source.tost1.equivalent\endcsname{0.0}
\expandafter\def\csname ppt@rev.desc.d1.uiv.p2.T.zoomclick-d3.tau-source.tost1.equivalent\endcsname{0.0}
\expandafter\def\csname val@rev.desc.d1.uiv.p2.T.zoomclick-d3.tau-source.tost1.p\endcsname{0.963}
\expandafter\def\csname cnt@rev.desc.d1.uiv.p2.T.zoomclick-d3.tau-source.tost1.p\endcsname{1}
\expandafter\def\csname pct@rev.desc.d1.uiv.p2.T.zoomclick-d3.tau-source.tost1.p\endcsname{96.3}
\expandafter\def\csname ppt@rev.desc.d1.uiv.p2.T.zoomclick-d3.tau-source.tost1.p\endcsname{$+$96.3}
\expandafter\def\csname val@rev.desc.d1.uiv.p2.T.zoomclick-d3.tau-source.tost5.ci90-inside\endcsname{0}
\expandafter\def\csname cnt@rev.desc.d1.uiv.p2.T.zoomclick-d3.tau-source.tost5.ci90-inside\endcsname{0}
\expandafter\def\csname pct@rev.desc.d1.uiv.p2.T.zoomclick-d3.tau-source.tost5.ci90-inside\endcsname{0.0}
\expandafter\def\csname ppt@rev.desc.d1.uiv.p2.T.zoomclick-d3.tau-source.tost5.ci90-inside\endcsname{0.0}
\expandafter\def\csname val@rev.desc.d1.uiv.p2.T.zoomclick-d3.tau-source.tost5.equivalent\endcsname{0}
\expandafter\def\csname cnt@rev.desc.d1.uiv.p2.T.zoomclick-d3.tau-source.tost5.equivalent\endcsname{0}
\expandafter\def\csname pct@rev.desc.d1.uiv.p2.T.zoomclick-d3.tau-source.tost5.equivalent\endcsname{0.0}
\expandafter\def\csname ppt@rev.desc.d1.uiv.p2.T.zoomclick-d3.tau-source.tost5.equivalent\endcsname{0.0}
\expandafter\def\csname val@rev.desc.d1.uiv.p2.T.zoomclick-d3.tau-source.tost5.p\endcsname{0.533}
\expandafter\def\csname cnt@rev.desc.d1.uiv.p2.T.zoomclick-d3.tau-source.tost5.p\endcsname{1}
\expandafter\def\csname pct@rev.desc.d1.uiv.p2.T.zoomclick-d3.tau-source.tost5.p\endcsname{53.3}
\expandafter\def\csname ppt@rev.desc.d1.uiv.p2.T.zoomclick-d3.tau-source.tost5.p\endcsname{$+$53.3}
\expandafter\def\csname val@rev.desc.d1.uiv.p2.T.zoomclick-d3.tau-target.ci90.0\endcsname{-0.042}
\expandafter\def\csname cnt@rev.desc.d1.uiv.p2.T.zoomclick-d3.tau-target.ci90.0\endcsname{0}
\expandafter\def\csname pct@rev.desc.d1.uiv.p2.T.zoomclick-d3.tau-target.ci90.0\endcsname{$-$4.2}
\expandafter\def\csname ppt@rev.desc.d1.uiv.p2.T.zoomclick-d3.tau-target.ci90.0\endcsname{$-$4.2}
\expandafter\def\csname val@rev.desc.d1.uiv.p2.T.zoomclick-d3.tau-target.ci90.1\endcsname{0.028}
\expandafter\def\csname cnt@rev.desc.d1.uiv.p2.T.zoomclick-d3.tau-target.ci90.1\endcsname{0}
\expandafter\def\csname pct@rev.desc.d1.uiv.p2.T.zoomclick-d3.tau-target.ci90.1\endcsname{2.8}
\expandafter\def\csname ppt@rev.desc.d1.uiv.p2.T.zoomclick-d3.tau-target.ci90.1\endcsname{$+$2.8}
\expandafter\def\csname val@rev.desc.d1.uiv.p2.T.zoomclick-d3.tau-target.ci95-excludes-0\endcsname{0}
\expandafter\def\csname cnt@rev.desc.d1.uiv.p2.T.zoomclick-d3.tau-target.ci95-excludes-0\endcsname{0}
\expandafter\def\csname pct@rev.desc.d1.uiv.p2.T.zoomclick-d3.tau-target.ci95-excludes-0\endcsname{0.0}
\expandafter\def\csname ppt@rev.desc.d1.uiv.p2.T.zoomclick-d3.tau-target.ci95-excludes-0\endcsname{0.0}
\expandafter\def\csname val@rev.desc.d1.uiv.p2.T.zoomclick-d3.tau-target.ci95.0\endcsname{-0.049}
\expandafter\def\csname cnt@rev.desc.d1.uiv.p2.T.zoomclick-d3.tau-target.ci95.0\endcsname{0}
\expandafter\def\csname pct@rev.desc.d1.uiv.p2.T.zoomclick-d3.tau-target.ci95.0\endcsname{$-$4.9}
\expandafter\def\csname ppt@rev.desc.d1.uiv.p2.T.zoomclick-d3.tau-target.ci95.0\endcsname{$-$4.9}
\expandafter\def\csname val@rev.desc.d1.uiv.p2.T.zoomclick-d3.tau-target.ci95.1\endcsname{0.028}
\expandafter\def\csname cnt@rev.desc.d1.uiv.p2.T.zoomclick-d3.tau-target.ci95.1\endcsname{0}
\expandafter\def\csname pct@rev.desc.d1.uiv.p2.T.zoomclick-d3.tau-target.ci95.1\endcsname{2.8}
\expandafter\def\csname ppt@rev.desc.d1.uiv.p2.T.zoomclick-d3.tau-target.ci95.1\endcsname{$+$2.8}
\expandafter\def\csname val@rev.desc.d1.uiv.p2.T.zoomclick-d3.tau-target.mean\endcsname{-0.007}
\expandafter\def\csname cnt@rev.desc.d1.uiv.p2.T.zoomclick-d3.tau-target.mean\endcsname{0}
\expandafter\def\csname pct@rev.desc.d1.uiv.p2.T.zoomclick-d3.tau-target.mean\endcsname{$-$0.7}
\expandafter\def\csname ppt@rev.desc.d1.uiv.p2.T.zoomclick-d3.tau-target.mean\endcsname{$-$0.7}
\expandafter\def\csname val@rev.desc.d1.uiv.p2.T.zoomclick-d3.tau-target.n-apps\endcsname{24}
\expandafter\def\csname cnt@rev.desc.d1.uiv.p2.T.zoomclick-d3.tau-target.n-apps\endcsname{24}
\expandafter\def\csname pct@rev.desc.d1.uiv.p2.T.zoomclick-d3.tau-target.n-apps\endcsname{2400.0}
\expandafter\def\csname ppt@rev.desc.d1.uiv.p2.T.zoomclick-d3.tau-target.n-apps\endcsname{$+$2400.0}
\expandafter\def\csname val@rev.desc.d1.uiv.p2.T.zoomclick-d3.tau-target.tost1.ci90-inside\endcsname{0}
\expandafter\def\csname cnt@rev.desc.d1.uiv.p2.T.zoomclick-d3.tau-target.tost1.ci90-inside\endcsname{0}
\expandafter\def\csname pct@rev.desc.d1.uiv.p2.T.zoomclick-d3.tau-target.tost1.ci90-inside\endcsname{0.0}
\expandafter\def\csname ppt@rev.desc.d1.uiv.p2.T.zoomclick-d3.tau-target.tost1.ci90-inside\endcsname{0.0}
\expandafter\def\csname val@rev.desc.d1.uiv.p2.T.zoomclick-d3.tau-target.tost1.equivalent\endcsname{0}
\expandafter\def\csname cnt@rev.desc.d1.uiv.p2.T.zoomclick-d3.tau-target.tost1.equivalent\endcsname{0}
\expandafter\def\csname pct@rev.desc.d1.uiv.p2.T.zoomclick-d3.tau-target.tost1.equivalent\endcsname{0.0}
\expandafter\def\csname ppt@rev.desc.d1.uiv.p2.T.zoomclick-d3.tau-target.tost1.equivalent\endcsname{0.0}
\expandafter\def\csname val@rev.desc.d1.uiv.p2.T.zoomclick-d3.tau-target.tost1.p\endcsname{0.431}
\expandafter\def\csname cnt@rev.desc.d1.uiv.p2.T.zoomclick-d3.tau-target.tost1.p\endcsname{0}
\expandafter\def\csname pct@rev.desc.d1.uiv.p2.T.zoomclick-d3.tau-target.tost1.p\endcsname{43.1}
\expandafter\def\csname ppt@rev.desc.d1.uiv.p2.T.zoomclick-d3.tau-target.tost1.p\endcsname{$+$43.1}
\expandafter\def\csname val@rev.desc.d1.uiv.p2.T.zoomclick-d3.tau-target.tost5.ci90-inside\endcsname{1}
\expandafter\def\csname cnt@rev.desc.d1.uiv.p2.T.zoomclick-d3.tau-target.tost5.ci90-inside\endcsname{1}
\expandafter\def\csname pct@rev.desc.d1.uiv.p2.T.zoomclick-d3.tau-target.tost5.ci90-inside\endcsname{100.0}
\expandafter\def\csname ppt@rev.desc.d1.uiv.p2.T.zoomclick-d3.tau-target.tost5.ci90-inside\endcsname{$+$100.0}
\expandafter\def\csname val@rev.desc.d1.uiv.p2.T.zoomclick-d3.tau-target.tost5.equivalent\endcsname{1}
\expandafter\def\csname cnt@rev.desc.d1.uiv.p2.T.zoomclick-d3.tau-target.tost5.equivalent\endcsname{1}
\expandafter\def\csname pct@rev.desc.d1.uiv.p2.T.zoomclick-d3.tau-target.tost5.equivalent\endcsname{100.0}
\expandafter\def\csname ppt@rev.desc.d1.uiv.p2.T.zoomclick-d3.tau-target.tost5.equivalent\endcsname{$+$100.0}
\expandafter\def\csname val@rev.desc.d1.uiv.p2.T.zoomclick-d3.tau-target.tost5.p\endcsname{0.021}
\expandafter\def\csname cnt@rev.desc.d1.uiv.p2.T.zoomclick-d3.tau-target.tost5.p\endcsname{0}
\expandafter\def\csname pct@rev.desc.d1.uiv.p2.T.zoomclick-d3.tau-target.tost5.p\endcsname{2.1}
\expandafter\def\csname ppt@rev.desc.d1.uiv.p2.T.zoomclick-d3.tau-target.tost5.p\endcsname{$+$2.1}
\expandafter\def\csname val@rev.desc.d1.uiv.p2.T.zoomclick-d3.zero-estimate\endcsname{0}
\expandafter\def\csname cnt@rev.desc.d1.uiv.p2.T.zoomclick-d3.zero-estimate\endcsname{0}
\expandafter\def\csname pct@rev.desc.d1.uiv.p2.T.zoomclick-d3.zero-estimate\endcsname{0.0}
\expandafter\def\csname ppt@rev.desc.d1.uiv.p2.T.zoomclick-d3.zero-estimate\endcsname{0.0}
\expandafter\def\csname val@rev.desc.d1.uiv.p2.Tsealed.center-d1.n-tasks\endcsname{144}
\expandafter\def\csname cnt@rev.desc.d1.uiv.p2.Tsealed.center-d1.n-tasks\endcsname{144}
\expandafter\def\csname pct@rev.desc.d1.uiv.p2.Tsealed.center-d1.n-tasks\endcsname{14400.0}
\expandafter\def\csname ppt@rev.desc.d1.uiv.p2.Tsealed.center-d1.n-tasks\endcsname{$+$14400.0}
\expandafter\def\csname val@rev.desc.d1.uiv.p2.Tsealed.center-d1.paper-tau-T\endcsname{-0.028}
\expandafter\def\csname cnt@rev.desc.d1.uiv.p2.Tsealed.center-d1.paper-tau-T\endcsname{0}
\expandafter\def\csname pct@rev.desc.d1.uiv.p2.Tsealed.center-d1.paper-tau-T\endcsname{$-$2.8}
\expandafter\def\csname ppt@rev.desc.d1.uiv.p2.Tsealed.center-d1.paper-tau-T\endcsname{$-$2.8}
\expandafter\def\csname val@rev.desc.d1.uiv.p2.Tsealed.center-d1.reproduces\endcsname{1}
\expandafter\def\csname cnt@rev.desc.d1.uiv.p2.Tsealed.center-d1.reproduces\endcsname{1}
\expandafter\def\csname pct@rev.desc.d1.uiv.p2.Tsealed.center-d1.reproduces\endcsname{100.0}
\expandafter\def\csname ppt@rev.desc.d1.uiv.p2.Tsealed.center-d1.reproduces\endcsname{$+$100.0}
\expandafter\def\csname val@rev.desc.d1.uiv.p2.Tsealed.center-d1.reproduces-ci\endcsname{1}
\expandafter\def\csname cnt@rev.desc.d1.uiv.p2.Tsealed.center-d1.reproduces-ci\endcsname{1}
\expandafter\def\csname pct@rev.desc.d1.uiv.p2.Tsealed.center-d1.reproduces-ci\endcsname{100.0}
\expandafter\def\csname ppt@rev.desc.d1.uiv.p2.Tsealed.center-d1.reproduces-ci\endcsname{$+$100.0}
\expandafter\def\csname val@rev.desc.d1.uiv.p2.Tsealed.center-d1.tau-target-T.ci90.0\endcsname{-0.062}
\expandafter\def\csname cnt@rev.desc.d1.uiv.p2.Tsealed.center-d1.tau-target-T.ci90.0\endcsname{0}
\expandafter\def\csname pct@rev.desc.d1.uiv.p2.Tsealed.center-d1.tau-target-T.ci90.0\endcsname{$-$6.2}
\expandafter\def\csname ppt@rev.desc.d1.uiv.p2.Tsealed.center-d1.tau-target-T.ci90.0\endcsname{$-$6.2}
\expandafter\def\csname val@rev.desc.d1.uiv.p2.Tsealed.center-d1.tau-target-T.ci90.1\endcsname{0.007}
\expandafter\def\csname cnt@rev.desc.d1.uiv.p2.Tsealed.center-d1.tau-target-T.ci90.1\endcsname{0}
\expandafter\def\csname pct@rev.desc.d1.uiv.p2.Tsealed.center-d1.tau-target-T.ci90.1\endcsname{0.7}
\expandafter\def\csname ppt@rev.desc.d1.uiv.p2.Tsealed.center-d1.tau-target-T.ci90.1\endcsname{$+$0.7}
\expandafter\def\csname val@rev.desc.d1.uiv.p2.Tsealed.center-d1.tau-target-T.ci95-excludes-0\endcsname{0}
\expandafter\def\csname cnt@rev.desc.d1.uiv.p2.Tsealed.center-d1.tau-target-T.ci95-excludes-0\endcsname{0}
\expandafter\def\csname pct@rev.desc.d1.uiv.p2.Tsealed.center-d1.tau-target-T.ci95-excludes-0\endcsname{0.0}
\expandafter\def\csname ppt@rev.desc.d1.uiv.p2.Tsealed.center-d1.tau-target-T.ci95-excludes-0\endcsname{0.0}
\expandafter\def\csname val@rev.desc.d1.uiv.p2.Tsealed.center-d1.tau-target-T.ci95.0\endcsname{-0.069}
\expandafter\def\csname cnt@rev.desc.d1.uiv.p2.Tsealed.center-d1.tau-target-T.ci95.0\endcsname{0}
\expandafter\def\csname pct@rev.desc.d1.uiv.p2.Tsealed.center-d1.tau-target-T.ci95.0\endcsname{$-$6.9}
\expandafter\def\csname ppt@rev.desc.d1.uiv.p2.Tsealed.center-d1.tau-target-T.ci95.0\endcsname{$-$6.9}
\expandafter\def\csname val@rev.desc.d1.uiv.p2.Tsealed.center-d1.tau-target-T.ci95.1\endcsname{0.007}
\expandafter\def\csname cnt@rev.desc.d1.uiv.p2.Tsealed.center-d1.tau-target-T.ci95.1\endcsname{0}
\expandafter\def\csname pct@rev.desc.d1.uiv.p2.Tsealed.center-d1.tau-target-T.ci95.1\endcsname{0.7}
\expandafter\def\csname ppt@rev.desc.d1.uiv.p2.Tsealed.center-d1.tau-target-T.ci95.1\endcsname{$+$0.7}
\expandafter\def\csname val@rev.desc.d1.uiv.p2.Tsealed.center-d1.tau-target-T.mean\endcsname{-0.028}
\expandafter\def\csname cnt@rev.desc.d1.uiv.p2.Tsealed.center-d1.tau-target-T.mean\endcsname{0}
\expandafter\def\csname pct@rev.desc.d1.uiv.p2.Tsealed.center-d1.tau-target-T.mean\endcsname{$-$2.8}
\expandafter\def\csname ppt@rev.desc.d1.uiv.p2.Tsealed.center-d1.tau-target-T.mean\endcsname{$-$2.8}
\expandafter\def\csname val@rev.desc.d1.uiv.p2.Tsealed.center-d1.tau-target-T.n-apps\endcsname{24}
\expandafter\def\csname cnt@rev.desc.d1.uiv.p2.Tsealed.center-d1.tau-target-T.n-apps\endcsname{24}
\expandafter\def\csname pct@rev.desc.d1.uiv.p2.Tsealed.center-d1.tau-target-T.n-apps\endcsname{2400.0}
\expandafter\def\csname ppt@rev.desc.d1.uiv.p2.Tsealed.center-d1.tau-target-T.n-apps\endcsname{$+$2400.0}
\expandafter\def\csname val@rev.desc.d1.uiv.p2.Tsealed.center-d1.tau-target-T.tost1.ci90-inside\endcsname{0}
\expandafter\def\csname cnt@rev.desc.d1.uiv.p2.Tsealed.center-d1.tau-target-T.tost1.ci90-inside\endcsname{0}
\expandafter\def\csname pct@rev.desc.d1.uiv.p2.Tsealed.center-d1.tau-target-T.tost1.ci90-inside\endcsname{0.0}
\expandafter\def\csname ppt@rev.desc.d1.uiv.p2.Tsealed.center-d1.tau-target-T.tost1.ci90-inside\endcsname{0.0}
\expandafter\def\csname val@rev.desc.d1.uiv.p2.Tsealed.center-d1.tau-target-T.tost1.equivalent\endcsname{0}
\expandafter\def\csname cnt@rev.desc.d1.uiv.p2.Tsealed.center-d1.tau-target-T.tost1.equivalent\endcsname{0}
\expandafter\def\csname pct@rev.desc.d1.uiv.p2.Tsealed.center-d1.tau-target-T.tost1.equivalent\endcsname{0.0}
\expandafter\def\csname ppt@rev.desc.d1.uiv.p2.Tsealed.center-d1.tau-target-T.tost1.equivalent\endcsname{0.0}
\expandafter\def\csname val@rev.desc.d1.uiv.p2.Tsealed.center-d1.tau-target-T.tost1.p\endcsname{0.785}
\expandafter\def\csname cnt@rev.desc.d1.uiv.p2.Tsealed.center-d1.tau-target-T.tost1.p\endcsname{1}
\expandafter\def\csname pct@rev.desc.d1.uiv.p2.Tsealed.center-d1.tau-target-T.tost1.p\endcsname{78.5}
\expandafter\def\csname ppt@rev.desc.d1.uiv.p2.Tsealed.center-d1.tau-target-T.tost1.p\endcsname{$+$78.5}
\expandafter\def\csname val@rev.desc.d1.uiv.p2.Tsealed.center-d1.tau-target-T.tost5.ci90-inside\endcsname{0}
\expandafter\def\csname cnt@rev.desc.d1.uiv.p2.Tsealed.center-d1.tau-target-T.tost5.ci90-inside\endcsname{0}
\expandafter\def\csname pct@rev.desc.d1.uiv.p2.Tsealed.center-d1.tau-target-T.tost5.ci90-inside\endcsname{0.0}
\expandafter\def\csname ppt@rev.desc.d1.uiv.p2.Tsealed.center-d1.tau-target-T.tost5.ci90-inside\endcsname{0.0}
\expandafter\def\csname val@rev.desc.d1.uiv.p2.Tsealed.center-d1.tau-target-T.tost5.equivalent\endcsname{0}
\expandafter\def\csname cnt@rev.desc.d1.uiv.p2.Tsealed.center-d1.tau-target-T.tost5.equivalent\endcsname{0}
\expandafter\def\csname pct@rev.desc.d1.uiv.p2.Tsealed.center-d1.tau-target-T.tost5.equivalent\endcsname{0.0}
\expandafter\def\csname ppt@rev.desc.d1.uiv.p2.Tsealed.center-d1.tau-target-T.tost5.equivalent\endcsname{0.0}
\expandafter\def\csname val@rev.desc.d1.uiv.p2.Tsealed.center-d1.tau-target-T.tost5.p\endcsname{0.134}
\expandafter\def\csname cnt@rev.desc.d1.uiv.p2.Tsealed.center-d1.tau-target-T.tost5.p\endcsname{0}
\expandafter\def\csname pct@rev.desc.d1.uiv.p2.Tsealed.center-d1.tau-target-T.tost5.p\endcsname{13.4}
\expandafter\def\csname ppt@rev.desc.d1.uiv.p2.Tsealed.center-d1.tau-target-T.tost5.p\endcsname{$+$13.4}
\expandafter\def\csname val@rev.desc.d1.uiv.p2.Tsealed.context-d1.n-tasks\endcsname{144}
\expandafter\def\csname cnt@rev.desc.d1.uiv.p2.Tsealed.context-d1.n-tasks\endcsname{144}
\expandafter\def\csname pct@rev.desc.d1.uiv.p2.Tsealed.context-d1.n-tasks\endcsname{14400.0}
\expandafter\def\csname ppt@rev.desc.d1.uiv.p2.Tsealed.context-d1.n-tasks\endcsname{$+$14400.0}
\expandafter\def\csname val@rev.desc.d1.uiv.p2.Tsealed.context-d1.paper-tau-T\endcsname{0.000}
\expandafter\def\csname cnt@rev.desc.d1.uiv.p2.Tsealed.context-d1.paper-tau-T\endcsname{0}
\expandafter\def\csname pct@rev.desc.d1.uiv.p2.Tsealed.context-d1.paper-tau-T\endcsname{0.0}
\expandafter\def\csname ppt@rev.desc.d1.uiv.p2.Tsealed.context-d1.paper-tau-T\endcsname{0.0}
\expandafter\def\csname val@rev.desc.d1.uiv.p2.Tsealed.context-d1.reproduces\endcsname{1}
\expandafter\def\csname cnt@rev.desc.d1.uiv.p2.Tsealed.context-d1.reproduces\endcsname{1}
\expandafter\def\csname pct@rev.desc.d1.uiv.p2.Tsealed.context-d1.reproduces\endcsname{100.0}
\expandafter\def\csname ppt@rev.desc.d1.uiv.p2.Tsealed.context-d1.reproduces\endcsname{$+$100.0}
\expandafter\def\csname val@rev.desc.d1.uiv.p2.Tsealed.context-d1.reproduces-ci\endcsname{1}
\expandafter\def\csname cnt@rev.desc.d1.uiv.p2.Tsealed.context-d1.reproduces-ci\endcsname{1}
\expandafter\def\csname pct@rev.desc.d1.uiv.p2.Tsealed.context-d1.reproduces-ci\endcsname{100.0}
\expandafter\def\csname ppt@rev.desc.d1.uiv.p2.Tsealed.context-d1.reproduces-ci\endcsname{$+$100.0}
\expandafter\def\csname val@rev.desc.d1.uiv.p2.Tsealed.context-d1.tau-target-T.ci90.0\endcsname{-0.021}
\expandafter\def\csname cnt@rev.desc.d1.uiv.p2.Tsealed.context-d1.tau-target-T.ci90.0\endcsname{0}
\expandafter\def\csname pct@rev.desc.d1.uiv.p2.Tsealed.context-d1.tau-target-T.ci90.0\endcsname{$-$2.1}
\expandafter\def\csname ppt@rev.desc.d1.uiv.p2.Tsealed.context-d1.tau-target-T.ci90.0\endcsname{$-$2.1}
\expandafter\def\csname val@rev.desc.d1.uiv.p2.Tsealed.context-d1.tau-target-T.ci90.1\endcsname{0.021}
\expandafter\def\csname cnt@rev.desc.d1.uiv.p2.Tsealed.context-d1.tau-target-T.ci90.1\endcsname{0}
\expandafter\def\csname pct@rev.desc.d1.uiv.p2.Tsealed.context-d1.tau-target-T.ci90.1\endcsname{2.1}
\expandafter\def\csname ppt@rev.desc.d1.uiv.p2.Tsealed.context-d1.tau-target-T.ci90.1\endcsname{$+$2.1}
\expandafter\def\csname val@rev.desc.d1.uiv.p2.Tsealed.context-d1.tau-target-T.ci95-excludes-0\endcsname{0}
\expandafter\def\csname cnt@rev.desc.d1.uiv.p2.Tsealed.context-d1.tau-target-T.ci95-excludes-0\endcsname{0}
\expandafter\def\csname pct@rev.desc.d1.uiv.p2.Tsealed.context-d1.tau-target-T.ci95-excludes-0\endcsname{0.0}
\expandafter\def\csname ppt@rev.desc.d1.uiv.p2.Tsealed.context-d1.tau-target-T.ci95-excludes-0\endcsname{0.0}
\expandafter\def\csname val@rev.desc.d1.uiv.p2.Tsealed.context-d1.tau-target-T.ci95.0\endcsname{-0.028}
\expandafter\def\csname cnt@rev.desc.d1.uiv.p2.Tsealed.context-d1.tau-target-T.ci95.0\endcsname{0}
\expandafter\def\csname pct@rev.desc.d1.uiv.p2.Tsealed.context-d1.tau-target-T.ci95.0\endcsname{$-$2.8}
\expandafter\def\csname ppt@rev.desc.d1.uiv.p2.Tsealed.context-d1.tau-target-T.ci95.0\endcsname{$-$2.8}
\expandafter\def\csname val@rev.desc.d1.uiv.p2.Tsealed.context-d1.tau-target-T.ci95.1\endcsname{0.028}
\expandafter\def\csname cnt@rev.desc.d1.uiv.p2.Tsealed.context-d1.tau-target-T.ci95.1\endcsname{0}
\expandafter\def\csname pct@rev.desc.d1.uiv.p2.Tsealed.context-d1.tau-target-T.ci95.1\endcsname{2.8}
\expandafter\def\csname ppt@rev.desc.d1.uiv.p2.Tsealed.context-d1.tau-target-T.ci95.1\endcsname{$+$2.8}
\expandafter\def\csname val@rev.desc.d1.uiv.p2.Tsealed.context-d1.tau-target-T.mean\endcsname{0.000}
\expandafter\def\csname cnt@rev.desc.d1.uiv.p2.Tsealed.context-d1.tau-target-T.mean\endcsname{0}
\expandafter\def\csname pct@rev.desc.d1.uiv.p2.Tsealed.context-d1.tau-target-T.mean\endcsname{0.0}
\expandafter\def\csname ppt@rev.desc.d1.uiv.p2.Tsealed.context-d1.tau-target-T.mean\endcsname{0.0}
\expandafter\def\csname val@rev.desc.d1.uiv.p2.Tsealed.context-d1.tau-target-T.n-apps\endcsname{24}
\expandafter\def\csname cnt@rev.desc.d1.uiv.p2.Tsealed.context-d1.tau-target-T.n-apps\endcsname{24}
\expandafter\def\csname pct@rev.desc.d1.uiv.p2.Tsealed.context-d1.tau-target-T.n-apps\endcsname{2400.0}
\expandafter\def\csname ppt@rev.desc.d1.uiv.p2.Tsealed.context-d1.tau-target-T.n-apps\endcsname{$+$2400.0}
\expandafter\def\csname val@rev.desc.d1.uiv.p2.Tsealed.context-d1.tau-target-T.tost1.ci90-inside\endcsname{0}
\expandafter\def\csname cnt@rev.desc.d1.uiv.p2.Tsealed.context-d1.tau-target-T.tost1.ci90-inside\endcsname{0}
\expandafter\def\csname pct@rev.desc.d1.uiv.p2.Tsealed.context-d1.tau-target-T.tost1.ci90-inside\endcsname{0.0}
\expandafter\def\csname ppt@rev.desc.d1.uiv.p2.Tsealed.context-d1.tau-target-T.tost1.ci90-inside\endcsname{0.0}
\expandafter\def\csname val@rev.desc.d1.uiv.p2.Tsealed.context-d1.tau-target-T.tost1.equivalent\endcsname{0}
\expandafter\def\csname cnt@rev.desc.d1.uiv.p2.Tsealed.context-d1.tau-target-T.tost1.equivalent\endcsname{0}
\expandafter\def\csname pct@rev.desc.d1.uiv.p2.Tsealed.context-d1.tau-target-T.tost1.equivalent\endcsname{0.0}
\expandafter\def\csname ppt@rev.desc.d1.uiv.p2.Tsealed.context-d1.tau-target-T.tost1.equivalent\endcsname{0.0}
\expandafter\def\csname val@rev.desc.d1.uiv.p2.Tsealed.context-d1.tau-target-T.tost1.p\endcsname{0.216}
\expandafter\def\csname cnt@rev.desc.d1.uiv.p2.Tsealed.context-d1.tau-target-T.tost1.p\endcsname{0}
\expandafter\def\csname pct@rev.desc.d1.uiv.p2.Tsealed.context-d1.tau-target-T.tost1.p\endcsname{21.6}
\expandafter\def\csname ppt@rev.desc.d1.uiv.p2.Tsealed.context-d1.tau-target-T.tost1.p\endcsname{$+$21.6}
\expandafter\def\csname val@rev.desc.d1.uiv.p2.Tsealed.context-d1.tau-target-T.tost5.ci90-inside\endcsname{1}
\expandafter\def\csname cnt@rev.desc.d1.uiv.p2.Tsealed.context-d1.tau-target-T.tost5.ci90-inside\endcsname{1}
\expandafter\def\csname pct@rev.desc.d1.uiv.p2.Tsealed.context-d1.tau-target-T.tost5.ci90-inside\endcsname{100.0}
\expandafter\def\csname ppt@rev.desc.d1.uiv.p2.Tsealed.context-d1.tau-target-T.tost5.ci90-inside\endcsname{$+$100.0}
\expandafter\def\csname val@rev.desc.d1.uiv.p2.Tsealed.context-d1.tau-target-T.tost5.equivalent\endcsname{1}
\expandafter\def\csname cnt@rev.desc.d1.uiv.p2.Tsealed.context-d1.tau-target-T.tost5.equivalent\endcsname{1}
\expandafter\def\csname pct@rev.desc.d1.uiv.p2.Tsealed.context-d1.tau-target-T.tost5.equivalent\endcsname{100.0}
\expandafter\def\csname ppt@rev.desc.d1.uiv.p2.Tsealed.context-d1.tau-target-T.tost5.equivalent\endcsname{$+$100.0}
\expandafter\def\csname val@rev.desc.d1.uiv.p2.Tsealed.context-d1.tau-target-T.tost5.p\endcsname{0.000}
\expandafter\def\csname cnt@rev.desc.d1.uiv.p2.Tsealed.context-d1.tau-target-T.tost5.p\endcsname{0}
\expandafter\def\csname pct@rev.desc.d1.uiv.p2.Tsealed.context-d1.tau-target-T.tost5.p\endcsname{0.0}
\expandafter\def\csname ppt@rev.desc.d1.uiv.p2.Tsealed.context-d1.tau-target-T.tost5.p\endcsname{0.0}
\expandafter\def\csname val@rev.desc.d1.uiv.p2.Tsealed.dual-d2.n-tasks\endcsname{144}
\expandafter\def\csname cnt@rev.desc.d1.uiv.p2.Tsealed.dual-d2.n-tasks\endcsname{144}
\expandafter\def\csname pct@rev.desc.d1.uiv.p2.Tsealed.dual-d2.n-tasks\endcsname{14400.0}
\expandafter\def\csname ppt@rev.desc.d1.uiv.p2.Tsealed.dual-d2.n-tasks\endcsname{$+$14400.0}
\expandafter\def\csname val@rev.desc.d1.uiv.p2.Tsealed.dual-d2.paper-tau-T\endcsname{-0.062}
\expandafter\def\csname cnt@rev.desc.d1.uiv.p2.Tsealed.dual-d2.paper-tau-T\endcsname{0}
\expandafter\def\csname pct@rev.desc.d1.uiv.p2.Tsealed.dual-d2.paper-tau-T\endcsname{$-$6.2}
\expandafter\def\csname ppt@rev.desc.d1.uiv.p2.Tsealed.dual-d2.paper-tau-T\endcsname{$-$6.2}
\expandafter\def\csname val@rev.desc.d1.uiv.p2.Tsealed.dual-d2.reproduces\endcsname{1}
\expandafter\def\csname cnt@rev.desc.d1.uiv.p2.Tsealed.dual-d2.reproduces\endcsname{1}
\expandafter\def\csname pct@rev.desc.d1.uiv.p2.Tsealed.dual-d2.reproduces\endcsname{100.0}
\expandafter\def\csname ppt@rev.desc.d1.uiv.p2.Tsealed.dual-d2.reproduces\endcsname{$+$100.0}
\expandafter\def\csname val@rev.desc.d1.uiv.p2.Tsealed.dual-d2.reproduces-ci\endcsname{1}
\expandafter\def\csname cnt@rev.desc.d1.uiv.p2.Tsealed.dual-d2.reproduces-ci\endcsname{1}
\expandafter\def\csname pct@rev.desc.d1.uiv.p2.Tsealed.dual-d2.reproduces-ci\endcsname{100.0}
\expandafter\def\csname ppt@rev.desc.d1.uiv.p2.Tsealed.dual-d2.reproduces-ci\endcsname{$+$100.0}
\expandafter\def\csname val@rev.desc.d1.uiv.p2.Tsealed.dual-d2.tau-target-T.ci90.0\endcsname{-0.111}
\expandafter\def\csname cnt@rev.desc.d1.uiv.p2.Tsealed.dual-d2.tau-target-T.ci90.0\endcsname{0}
\expandafter\def\csname pct@rev.desc.d1.uiv.p2.Tsealed.dual-d2.tau-target-T.ci90.0\endcsname{$-$11.1}
\expandafter\def\csname ppt@rev.desc.d1.uiv.p2.Tsealed.dual-d2.tau-target-T.ci90.0\endcsname{$-$11.1}
\expandafter\def\csname val@rev.desc.d1.uiv.p2.Tsealed.dual-d2.tau-target-T.ci90.1\endcsname{-0.014}
\expandafter\def\csname cnt@rev.desc.d1.uiv.p2.Tsealed.dual-d2.tau-target-T.ci90.1\endcsname{0}
\expandafter\def\csname pct@rev.desc.d1.uiv.p2.Tsealed.dual-d2.tau-target-T.ci90.1\endcsname{$-$1.4}
\expandafter\def\csname ppt@rev.desc.d1.uiv.p2.Tsealed.dual-d2.tau-target-T.ci90.1\endcsname{$-$1.4}
\expandafter\def\csname val@rev.desc.d1.uiv.p2.Tsealed.dual-d2.tau-target-T.ci95-excludes-0\endcsname{1}
\expandafter\def\csname cnt@rev.desc.d1.uiv.p2.Tsealed.dual-d2.tau-target-T.ci95-excludes-0\endcsname{1}
\expandafter\def\csname pct@rev.desc.d1.uiv.p2.Tsealed.dual-d2.tau-target-T.ci95-excludes-0\endcsname{100.0}
\expandafter\def\csname ppt@rev.desc.d1.uiv.p2.Tsealed.dual-d2.tau-target-T.ci95-excludes-0\endcsname{$+$100.0}
\expandafter\def\csname val@rev.desc.d1.uiv.p2.Tsealed.dual-d2.tau-target-T.ci95.0\endcsname{-0.125}
\expandafter\def\csname cnt@rev.desc.d1.uiv.p2.Tsealed.dual-d2.tau-target-T.ci95.0\endcsname{0}
\expandafter\def\csname pct@rev.desc.d1.uiv.p2.Tsealed.dual-d2.tau-target-T.ci95.0\endcsname{$-$12.5}
\expandafter\def\csname ppt@rev.desc.d1.uiv.p2.Tsealed.dual-d2.tau-target-T.ci95.0\endcsname{$-$12.5}
\expandafter\def\csname val@rev.desc.d1.uiv.p2.Tsealed.dual-d2.tau-target-T.ci95.1\endcsname{-0.007}
\expandafter\def\csname cnt@rev.desc.d1.uiv.p2.Tsealed.dual-d2.tau-target-T.ci95.1\endcsname{0}
\expandafter\def\csname pct@rev.desc.d1.uiv.p2.Tsealed.dual-d2.tau-target-T.ci95.1\endcsname{$-$0.7}
\expandafter\def\csname ppt@rev.desc.d1.uiv.p2.Tsealed.dual-d2.tau-target-T.ci95.1\endcsname{$-$0.7}
\expandafter\def\csname val@rev.desc.d1.uiv.p2.Tsealed.dual-d2.tau-target-T.mean\endcsname{-0.062}
\expandafter\def\csname cnt@rev.desc.d1.uiv.p2.Tsealed.dual-d2.tau-target-T.mean\endcsname{0}
\expandafter\def\csname pct@rev.desc.d1.uiv.p2.Tsealed.dual-d2.tau-target-T.mean\endcsname{$-$6.2}
\expandafter\def\csname ppt@rev.desc.d1.uiv.p2.Tsealed.dual-d2.tau-target-T.mean\endcsname{$-$6.2}
\expandafter\def\csname val@rev.desc.d1.uiv.p2.Tsealed.dual-d2.tau-target-T.n-apps\endcsname{24}
\expandafter\def\csname cnt@rev.desc.d1.uiv.p2.Tsealed.dual-d2.tau-target-T.n-apps\endcsname{24}
\expandafter\def\csname pct@rev.desc.d1.uiv.p2.Tsealed.dual-d2.tau-target-T.n-apps\endcsname{2400.0}
\expandafter\def\csname ppt@rev.desc.d1.uiv.p2.Tsealed.dual-d2.tau-target-T.n-apps\endcsname{$+$2400.0}
\expandafter\def\csname val@rev.desc.d1.uiv.p2.Tsealed.dual-d2.tau-target-T.tost1.ci90-inside\endcsname{0}
\expandafter\def\csname cnt@rev.desc.d1.uiv.p2.Tsealed.dual-d2.tau-target-T.tost1.ci90-inside\endcsname{0}
\expandafter\def\csname pct@rev.desc.d1.uiv.p2.Tsealed.dual-d2.tau-target-T.tost1.ci90-inside\endcsname{0.0}
\expandafter\def\csname ppt@rev.desc.d1.uiv.p2.Tsealed.dual-d2.tau-target-T.tost1.ci90-inside\endcsname{0.0}
\expandafter\def\csname val@rev.desc.d1.uiv.p2.Tsealed.dual-d2.tau-target-T.tost1.equivalent\endcsname{0}
\expandafter\def\csname cnt@rev.desc.d1.uiv.p2.Tsealed.dual-d2.tau-target-T.tost1.equivalent\endcsname{0}
\expandafter\def\csname pct@rev.desc.d1.uiv.p2.Tsealed.dual-d2.tau-target-T.tost1.equivalent\endcsname{0.0}
\expandafter\def\csname ppt@rev.desc.d1.uiv.p2.Tsealed.dual-d2.tau-target-T.tost1.equivalent\endcsname{0.0}
\expandafter\def\csname val@rev.desc.d1.uiv.p2.Tsealed.dual-d2.tau-target-T.tost1.p\endcsname{0.965}
\expandafter\def\csname cnt@rev.desc.d1.uiv.p2.Tsealed.dual-d2.tau-target-T.tost1.p\endcsname{1}
\expandafter\def\csname pct@rev.desc.d1.uiv.p2.Tsealed.dual-d2.tau-target-T.tost1.p\endcsname{96.5}
\expandafter\def\csname ppt@rev.desc.d1.uiv.p2.Tsealed.dual-d2.tau-target-T.tost1.p\endcsname{$+$96.5}
\expandafter\def\csname val@rev.desc.d1.uiv.p2.Tsealed.dual-d2.tau-target-T.tost5.ci90-inside\endcsname{0}
\expandafter\def\csname cnt@rev.desc.d1.uiv.p2.Tsealed.dual-d2.tau-target-T.tost5.ci90-inside\endcsname{0}
\expandafter\def\csname pct@rev.desc.d1.uiv.p2.Tsealed.dual-d2.tau-target-T.tost5.ci90-inside\endcsname{0.0}
\expandafter\def\csname ppt@rev.desc.d1.uiv.p2.Tsealed.dual-d2.tau-target-T.tost5.ci90-inside\endcsname{0.0}
\expandafter\def\csname val@rev.desc.d1.uiv.p2.Tsealed.dual-d2.tau-target-T.tost5.equivalent\endcsname{0}
\expandafter\def\csname cnt@rev.desc.d1.uiv.p2.Tsealed.dual-d2.tau-target-T.tost5.equivalent\endcsname{0}
\expandafter\def\csname pct@rev.desc.d1.uiv.p2.Tsealed.dual-d2.tau-target-T.tost5.equivalent\endcsname{0.0}
\expandafter\def\csname ppt@rev.desc.d1.uiv.p2.Tsealed.dual-d2.tau-target-T.tost5.equivalent\endcsname{0.0}
\expandafter\def\csname val@rev.desc.d1.uiv.p2.Tsealed.dual-d2.tau-target-T.tost5.p\endcsname{0.614}
\expandafter\def\csname cnt@rev.desc.d1.uiv.p2.Tsealed.dual-d2.tau-target-T.tost5.p\endcsname{1}
\expandafter\def\csname pct@rev.desc.d1.uiv.p2.Tsealed.dual-d2.tau-target-T.tost5.p\endcsname{61.4}
\expandafter\def\csname ppt@rev.desc.d1.uiv.p2.Tsealed.dual-d2.tau-target-T.tost5.p\endcsname{$+$61.4}
\expandafter\def\csname val@rev.desc.d1.uiv.p2.Tsealed.grid2-d3.n-tasks\endcsname{144}
\expandafter\def\csname cnt@rev.desc.d1.uiv.p2.Tsealed.grid2-d3.n-tasks\endcsname{144}
\expandafter\def\csname pct@rev.desc.d1.uiv.p2.Tsealed.grid2-d3.n-tasks\endcsname{14400.0}
\expandafter\def\csname ppt@rev.desc.d1.uiv.p2.Tsealed.grid2-d3.n-tasks\endcsname{$+$14400.0}
\expandafter\def\csname val@rev.desc.d1.uiv.p2.Tsealed.grid2-d3.paper-tau-T\endcsname{-0.007}
\expandafter\def\csname cnt@rev.desc.d1.uiv.p2.Tsealed.grid2-d3.paper-tau-T\endcsname{0}
\expandafter\def\csname pct@rev.desc.d1.uiv.p2.Tsealed.grid2-d3.paper-tau-T\endcsname{$-$0.7}
\expandafter\def\csname ppt@rev.desc.d1.uiv.p2.Tsealed.grid2-d3.paper-tau-T\endcsname{$-$0.7}
\expandafter\def\csname val@rev.desc.d1.uiv.p2.Tsealed.grid2-d3.reproduces\endcsname{1}
\expandafter\def\csname cnt@rev.desc.d1.uiv.p2.Tsealed.grid2-d3.reproduces\endcsname{1}
\expandafter\def\csname pct@rev.desc.d1.uiv.p2.Tsealed.grid2-d3.reproduces\endcsname{100.0}
\expandafter\def\csname ppt@rev.desc.d1.uiv.p2.Tsealed.grid2-d3.reproduces\endcsname{$+$100.0}
\expandafter\def\csname val@rev.desc.d1.uiv.p2.Tsealed.grid2-d3.reproduces-ci\endcsname{1}
\expandafter\def\csname cnt@rev.desc.d1.uiv.p2.Tsealed.grid2-d3.reproduces-ci\endcsname{1}
\expandafter\def\csname pct@rev.desc.d1.uiv.p2.Tsealed.grid2-d3.reproduces-ci\endcsname{100.0}
\expandafter\def\csname ppt@rev.desc.d1.uiv.p2.Tsealed.grid2-d3.reproduces-ci\endcsname{$+$100.0}
\expandafter\def\csname val@rev.desc.d1.uiv.p2.Tsealed.grid2-d3.tau-target-T.ci90.0\endcsname{-0.035}
\expandafter\def\csname cnt@rev.desc.d1.uiv.p2.Tsealed.grid2-d3.tau-target-T.ci90.0\endcsname{0}
\expandafter\def\csname pct@rev.desc.d1.uiv.p2.Tsealed.grid2-d3.tau-target-T.ci90.0\endcsname{$-$3.5}
\expandafter\def\csname ppt@rev.desc.d1.uiv.p2.Tsealed.grid2-d3.tau-target-T.ci90.0\endcsname{$-$3.5}
\expandafter\def\csname val@rev.desc.d1.uiv.p2.Tsealed.grid2-d3.tau-target-T.ci90.1\endcsname{0.021}
\expandafter\def\csname cnt@rev.desc.d1.uiv.p2.Tsealed.grid2-d3.tau-target-T.ci90.1\endcsname{0}
\expandafter\def\csname pct@rev.desc.d1.uiv.p2.Tsealed.grid2-d3.tau-target-T.ci90.1\endcsname{2.1}
\expandafter\def\csname ppt@rev.desc.d1.uiv.p2.Tsealed.grid2-d3.tau-target-T.ci90.1\endcsname{$+$2.1}
\expandafter\def\csname val@rev.desc.d1.uiv.p2.Tsealed.grid2-d3.tau-target-T.ci95-excludes-0\endcsname{0}
\expandafter\def\csname cnt@rev.desc.d1.uiv.p2.Tsealed.grid2-d3.tau-target-T.ci95-excludes-0\endcsname{0}
\expandafter\def\csname pct@rev.desc.d1.uiv.p2.Tsealed.grid2-d3.tau-target-T.ci95-excludes-0\endcsname{0.0}
\expandafter\def\csname ppt@rev.desc.d1.uiv.p2.Tsealed.grid2-d3.tau-target-T.ci95-excludes-0\endcsname{0.0}
\expandafter\def\csname val@rev.desc.d1.uiv.p2.Tsealed.grid2-d3.tau-target-T.ci95.0\endcsname{-0.042}
\expandafter\def\csname cnt@rev.desc.d1.uiv.p2.Tsealed.grid2-d3.tau-target-T.ci95.0\endcsname{0}
\expandafter\def\csname pct@rev.desc.d1.uiv.p2.Tsealed.grid2-d3.tau-target-T.ci95.0\endcsname{$-$4.2}
\expandafter\def\csname ppt@rev.desc.d1.uiv.p2.Tsealed.grid2-d3.tau-target-T.ci95.0\endcsname{$-$4.2}
\expandafter\def\csname val@rev.desc.d1.uiv.p2.Tsealed.grid2-d3.tau-target-T.ci95.1\endcsname{0.028}
\expandafter\def\csname cnt@rev.desc.d1.uiv.p2.Tsealed.grid2-d3.tau-target-T.ci95.1\endcsname{0}
\expandafter\def\csname pct@rev.desc.d1.uiv.p2.Tsealed.grid2-d3.tau-target-T.ci95.1\endcsname{2.8}
\expandafter\def\csname ppt@rev.desc.d1.uiv.p2.Tsealed.grid2-d3.tau-target-T.ci95.1\endcsname{$+$2.8}
\expandafter\def\csname val@rev.desc.d1.uiv.p2.Tsealed.grid2-d3.tau-target-T.mean\endcsname{-0.007}
\expandafter\def\csname cnt@rev.desc.d1.uiv.p2.Tsealed.grid2-d3.tau-target-T.mean\endcsname{0}
\expandafter\def\csname pct@rev.desc.d1.uiv.p2.Tsealed.grid2-d3.tau-target-T.mean\endcsname{$-$0.7}
\expandafter\def\csname ppt@rev.desc.d1.uiv.p2.Tsealed.grid2-d3.tau-target-T.mean\endcsname{$-$0.7}
\expandafter\def\csname val@rev.desc.d1.uiv.p2.Tsealed.grid2-d3.tau-target-T.n-apps\endcsname{24}
\expandafter\def\csname cnt@rev.desc.d1.uiv.p2.Tsealed.grid2-d3.tau-target-T.n-apps\endcsname{24}
\expandafter\def\csname pct@rev.desc.d1.uiv.p2.Tsealed.grid2-d3.tau-target-T.n-apps\endcsname{2400.0}
\expandafter\def\csname ppt@rev.desc.d1.uiv.p2.Tsealed.grid2-d3.tau-target-T.n-apps\endcsname{$+$2400.0}
\expandafter\def\csname val@rev.desc.d1.uiv.p2.Tsealed.grid2-d3.tau-target-T.tost1.ci90-inside\endcsname{0}
\expandafter\def\csname cnt@rev.desc.d1.uiv.p2.Tsealed.grid2-d3.tau-target-T.tost1.ci90-inside\endcsname{0}
\expandafter\def\csname pct@rev.desc.d1.uiv.p2.Tsealed.grid2-d3.tau-target-T.tost1.ci90-inside\endcsname{0.0}
\expandafter\def\csname ppt@rev.desc.d1.uiv.p2.Tsealed.grid2-d3.tau-target-T.tost1.ci90-inside\endcsname{0.0}
\expandafter\def\csname val@rev.desc.d1.uiv.p2.Tsealed.grid2-d3.tau-target-T.tost1.equivalent\endcsname{0}
\expandafter\def\csname cnt@rev.desc.d1.uiv.p2.Tsealed.grid2-d3.tau-target-T.tost1.equivalent\endcsname{0}
\expandafter\def\csname pct@rev.desc.d1.uiv.p2.Tsealed.grid2-d3.tau-target-T.tost1.equivalent\endcsname{0.0}
\expandafter\def\csname ppt@rev.desc.d1.uiv.p2.Tsealed.grid2-d3.tau-target-T.tost1.equivalent\endcsname{0.0}
\expandafter\def\csname val@rev.desc.d1.uiv.p2.Tsealed.grid2-d3.tau-target-T.tost1.p\endcsname{0.430}
\expandafter\def\csname cnt@rev.desc.d1.uiv.p2.Tsealed.grid2-d3.tau-target-T.tost1.p\endcsname{0}
\expandafter\def\csname pct@rev.desc.d1.uiv.p2.Tsealed.grid2-d3.tau-target-T.tost1.p\endcsname{43.0}
\expandafter\def\csname ppt@rev.desc.d1.uiv.p2.Tsealed.grid2-d3.tau-target-T.tost1.p\endcsname{$+$43.0}
\expandafter\def\csname val@rev.desc.d1.uiv.p2.Tsealed.grid2-d3.tau-target-T.tost5.ci90-inside\endcsname{1}
\expandafter\def\csname cnt@rev.desc.d1.uiv.p2.Tsealed.grid2-d3.tau-target-T.tost5.ci90-inside\endcsname{1}
\expandafter\def\csname pct@rev.desc.d1.uiv.p2.Tsealed.grid2-d3.tau-target-T.tost5.ci90-inside\endcsname{100.0}
\expandafter\def\csname ppt@rev.desc.d1.uiv.p2.Tsealed.grid2-d3.tau-target-T.tost5.ci90-inside\endcsname{$+$100.0}
\expandafter\def\csname val@rev.desc.d1.uiv.p2.Tsealed.grid2-d3.tau-target-T.tost5.equivalent\endcsname{1}
\expandafter\def\csname cnt@rev.desc.d1.uiv.p2.Tsealed.grid2-d3.tau-target-T.tost5.equivalent\endcsname{1}
\expandafter\def\csname pct@rev.desc.d1.uiv.p2.Tsealed.grid2-d3.tau-target-T.tost5.equivalent\endcsname{100.0}
\expandafter\def\csname ppt@rev.desc.d1.uiv.p2.Tsealed.grid2-d3.tau-target-T.tost5.equivalent\endcsname{$+$100.0}
\expandafter\def\csname val@rev.desc.d1.uiv.p2.Tsealed.grid2-d3.tau-target-T.tost5.p\endcsname{0.006}
\expandafter\def\csname cnt@rev.desc.d1.uiv.p2.Tsealed.grid2-d3.tau-target-T.tost5.p\endcsname{0}
\expandafter\def\csname pct@rev.desc.d1.uiv.p2.Tsealed.grid2-d3.tau-target-T.tost5.p\endcsname{0.6}
\expandafter\def\csname ppt@rev.desc.d1.uiv.p2.Tsealed.grid2-d3.tau-target-T.tost5.p\endcsname{$+$0.6}
\expandafter\def\csname val@rev.desc.d1.uiv.p2.Tsealed.zoomclick-d3.n-tasks\endcsname{144}
\expandafter\def\csname cnt@rev.desc.d1.uiv.p2.Tsealed.zoomclick-d3.n-tasks\endcsname{144}
\expandafter\def\csname pct@rev.desc.d1.uiv.p2.Tsealed.zoomclick-d3.n-tasks\endcsname{14400.0}
\expandafter\def\csname ppt@rev.desc.d1.uiv.p2.Tsealed.zoomclick-d3.n-tasks\endcsname{$+$14400.0}
\expandafter\def\csname val@rev.desc.d1.uiv.p2.Tsealed.zoomclick-d3.paper-tau-T\endcsname{-0.007}
\expandafter\def\csname cnt@rev.desc.d1.uiv.p2.Tsealed.zoomclick-d3.paper-tau-T\endcsname{0}
\expandafter\def\csname pct@rev.desc.d1.uiv.p2.Tsealed.zoomclick-d3.paper-tau-T\endcsname{$-$0.7}
\expandafter\def\csname ppt@rev.desc.d1.uiv.p2.Tsealed.zoomclick-d3.paper-tau-T\endcsname{$-$0.7}
\expandafter\def\csname val@rev.desc.d1.uiv.p2.Tsealed.zoomclick-d3.reproduces\endcsname{1}
\expandafter\def\csname cnt@rev.desc.d1.uiv.p2.Tsealed.zoomclick-d3.reproduces\endcsname{1}
\expandafter\def\csname pct@rev.desc.d1.uiv.p2.Tsealed.zoomclick-d3.reproduces\endcsname{100.0}
\expandafter\def\csname ppt@rev.desc.d1.uiv.p2.Tsealed.zoomclick-d3.reproduces\endcsname{$+$100.0}
\expandafter\def\csname val@rev.desc.d1.uiv.p2.Tsealed.zoomclick-d3.reproduces-ci\endcsname{1}
\expandafter\def\csname cnt@rev.desc.d1.uiv.p2.Tsealed.zoomclick-d3.reproduces-ci\endcsname{1}
\expandafter\def\csname pct@rev.desc.d1.uiv.p2.Tsealed.zoomclick-d3.reproduces-ci\endcsname{100.0}
\expandafter\def\csname ppt@rev.desc.d1.uiv.p2.Tsealed.zoomclick-d3.reproduces-ci\endcsname{$+$100.0}
\expandafter\def\csname val@rev.desc.d1.uiv.p2.Tsealed.zoomclick-d3.tau-target-T.ci90.0\endcsname{-0.042}
\expandafter\def\csname cnt@rev.desc.d1.uiv.p2.Tsealed.zoomclick-d3.tau-target-T.ci90.0\endcsname{0}
\expandafter\def\csname pct@rev.desc.d1.uiv.p2.Tsealed.zoomclick-d3.tau-target-T.ci90.0\endcsname{$-$4.2}
\expandafter\def\csname ppt@rev.desc.d1.uiv.p2.Tsealed.zoomclick-d3.tau-target-T.ci90.0\endcsname{$-$4.2}
\expandafter\def\csname val@rev.desc.d1.uiv.p2.Tsealed.zoomclick-d3.tau-target-T.ci90.1\endcsname{0.028}
\expandafter\def\csname cnt@rev.desc.d1.uiv.p2.Tsealed.zoomclick-d3.tau-target-T.ci90.1\endcsname{0}
\expandafter\def\csname pct@rev.desc.d1.uiv.p2.Tsealed.zoomclick-d3.tau-target-T.ci90.1\endcsname{2.8}
\expandafter\def\csname ppt@rev.desc.d1.uiv.p2.Tsealed.zoomclick-d3.tau-target-T.ci90.1\endcsname{$+$2.8}
\expandafter\def\csname val@rev.desc.d1.uiv.p2.Tsealed.zoomclick-d3.tau-target-T.ci95-excludes-0\endcsname{0}
\expandafter\def\csname cnt@rev.desc.d1.uiv.p2.Tsealed.zoomclick-d3.tau-target-T.ci95-excludes-0\endcsname{0}
\expandafter\def\csname pct@rev.desc.d1.uiv.p2.Tsealed.zoomclick-d3.tau-target-T.ci95-excludes-0\endcsname{0.0}
\expandafter\def\csname ppt@rev.desc.d1.uiv.p2.Tsealed.zoomclick-d3.tau-target-T.ci95-excludes-0\endcsname{0.0}
\expandafter\def\csname val@rev.desc.d1.uiv.p2.Tsealed.zoomclick-d3.tau-target-T.ci95.0\endcsname{-0.049}
\expandafter\def\csname cnt@rev.desc.d1.uiv.p2.Tsealed.zoomclick-d3.tau-target-T.ci95.0\endcsname{0}
\expandafter\def\csname pct@rev.desc.d1.uiv.p2.Tsealed.zoomclick-d3.tau-target-T.ci95.0\endcsname{$-$4.9}
\expandafter\def\csname ppt@rev.desc.d1.uiv.p2.Tsealed.zoomclick-d3.tau-target-T.ci95.0\endcsname{$-$4.9}
\expandafter\def\csname val@rev.desc.d1.uiv.p2.Tsealed.zoomclick-d3.tau-target-T.ci95.1\endcsname{0.028}
\expandafter\def\csname cnt@rev.desc.d1.uiv.p2.Tsealed.zoomclick-d3.tau-target-T.ci95.1\endcsname{0}
\expandafter\def\csname pct@rev.desc.d1.uiv.p2.Tsealed.zoomclick-d3.tau-target-T.ci95.1\endcsname{2.8}
\expandafter\def\csname ppt@rev.desc.d1.uiv.p2.Tsealed.zoomclick-d3.tau-target-T.ci95.1\endcsname{$+$2.8}
\expandafter\def\csname val@rev.desc.d1.uiv.p2.Tsealed.zoomclick-d3.tau-target-T.mean\endcsname{-0.007}
\expandafter\def\csname cnt@rev.desc.d1.uiv.p2.Tsealed.zoomclick-d3.tau-target-T.mean\endcsname{0}
\expandafter\def\csname pct@rev.desc.d1.uiv.p2.Tsealed.zoomclick-d3.tau-target-T.mean\endcsname{$-$0.7}
\expandafter\def\csname ppt@rev.desc.d1.uiv.p2.Tsealed.zoomclick-d3.tau-target-T.mean\endcsname{$-$0.7}
\expandafter\def\csname val@rev.desc.d1.uiv.p2.Tsealed.zoomclick-d3.tau-target-T.n-apps\endcsname{24}
\expandafter\def\csname cnt@rev.desc.d1.uiv.p2.Tsealed.zoomclick-d3.tau-target-T.n-apps\endcsname{24}
\expandafter\def\csname pct@rev.desc.d1.uiv.p2.Tsealed.zoomclick-d3.tau-target-T.n-apps\endcsname{2400.0}
\expandafter\def\csname ppt@rev.desc.d1.uiv.p2.Tsealed.zoomclick-d3.tau-target-T.n-apps\endcsname{$+$2400.0}
\expandafter\def\csname val@rev.desc.d1.uiv.p2.Tsealed.zoomclick-d3.tau-target-T.tost1.ci90-inside\endcsname{0}
\expandafter\def\csname cnt@rev.desc.d1.uiv.p2.Tsealed.zoomclick-d3.tau-target-T.tost1.ci90-inside\endcsname{0}
\expandafter\def\csname pct@rev.desc.d1.uiv.p2.Tsealed.zoomclick-d3.tau-target-T.tost1.ci90-inside\endcsname{0.0}
\expandafter\def\csname ppt@rev.desc.d1.uiv.p2.Tsealed.zoomclick-d3.tau-target-T.tost1.ci90-inside\endcsname{0.0}
\expandafter\def\csname val@rev.desc.d1.uiv.p2.Tsealed.zoomclick-d3.tau-target-T.tost1.equivalent\endcsname{0}
\expandafter\def\csname cnt@rev.desc.d1.uiv.p2.Tsealed.zoomclick-d3.tau-target-T.tost1.equivalent\endcsname{0}
\expandafter\def\csname pct@rev.desc.d1.uiv.p2.Tsealed.zoomclick-d3.tau-target-T.tost1.equivalent\endcsname{0.0}
\expandafter\def\csname ppt@rev.desc.d1.uiv.p2.Tsealed.zoomclick-d3.tau-target-T.tost1.equivalent\endcsname{0.0}
\expandafter\def\csname val@rev.desc.d1.uiv.p2.Tsealed.zoomclick-d3.tau-target-T.tost1.p\endcsname{0.431}
\expandafter\def\csname cnt@rev.desc.d1.uiv.p2.Tsealed.zoomclick-d3.tau-target-T.tost1.p\endcsname{0}
\expandafter\def\csname pct@rev.desc.d1.uiv.p2.Tsealed.zoomclick-d3.tau-target-T.tost1.p\endcsname{43.1}
\expandafter\def\csname ppt@rev.desc.d1.uiv.p2.Tsealed.zoomclick-d3.tau-target-T.tost1.p\endcsname{$+$43.1}
\expandafter\def\csname val@rev.desc.d1.uiv.p2.Tsealed.zoomclick-d3.tau-target-T.tost5.ci90-inside\endcsname{1}
\expandafter\def\csname cnt@rev.desc.d1.uiv.p2.Tsealed.zoomclick-d3.tau-target-T.tost5.ci90-inside\endcsname{1}
\expandafter\def\csname pct@rev.desc.d1.uiv.p2.Tsealed.zoomclick-d3.tau-target-T.tost5.ci90-inside\endcsname{100.0}
\expandafter\def\csname ppt@rev.desc.d1.uiv.p2.Tsealed.zoomclick-d3.tau-target-T.tost5.ci90-inside\endcsname{$+$100.0}
\expandafter\def\csname val@rev.desc.d1.uiv.p2.Tsealed.zoomclick-d3.tau-target-T.tost5.equivalent\endcsname{1}
\expandafter\def\csname cnt@rev.desc.d1.uiv.p2.Tsealed.zoomclick-d3.tau-target-T.tost5.equivalent\endcsname{1}
\expandafter\def\csname pct@rev.desc.d1.uiv.p2.Tsealed.zoomclick-d3.tau-target-T.tost5.equivalent\endcsname{100.0}
\expandafter\def\csname ppt@rev.desc.d1.uiv.p2.Tsealed.zoomclick-d3.tau-target-T.tost5.equivalent\endcsname{$+$100.0}
\expandafter\def\csname val@rev.desc.d1.uiv.p2.Tsealed.zoomclick-d3.tau-target-T.tost5.p\endcsname{0.021}
\expandafter\def\csname cnt@rev.desc.d1.uiv.p2.Tsealed.zoomclick-d3.tau-target-T.tost5.p\endcsname{0}
\expandafter\def\csname pct@rev.desc.d1.uiv.p2.Tsealed.zoomclick-d3.tau-target-T.tost5.p\endcsname{2.1}
\expandafter\def\csname ppt@rev.desc.d1.uiv.p2.Tsealed.zoomclick-d3.tau-target-T.tost5.p\endcsname{$+$2.1}
\expandafter\def\csname val@rev.desc.d2.sspro.p1.C.D.acc-a\endcsname{0.562}
\expandafter\def\csname cnt@rev.desc.d2.sspro.p1.C.D.acc-a\endcsname{1}
\expandafter\def\csname pct@rev.desc.d2.sspro.p1.C.D.acc-a\endcsname{56.2}
\expandafter\def\csname ppt@rev.desc.d2.sspro.p1.C.D.acc-a\endcsname{$+$56.2}
\expandafter\def\csname val@rev.desc.d2.sspro.p1.C.D.acc-b\endcsname{0.646}
\expandafter\def\csname cnt@rev.desc.d2.sspro.p1.C.D.acc-b\endcsname{1}
\expandafter\def\csname pct@rev.desc.d2.sspro.p1.C.D.acc-b\endcsname{64.6}
\expandafter\def\csname ppt@rev.desc.d2.sspro.p1.C.D.acc-b\endcsname{$+$64.6}
\expandafter\def\csname val@rev.desc.d2.sspro.p1.C.D.cond01\endcsname{0.286}
\expandafter\def\csname cnt@rev.desc.d2.sspro.p1.C.D.cond01\endcsname{0}
\expandafter\def\csname pct@rev.desc.d2.sspro.p1.C.D.cond01\endcsname{28.6}
\expandafter\def\csname ppt@rev.desc.d2.sspro.p1.C.D.cond01\endcsname{$+$28.6}
\expandafter\def\csname val@rev.desc.d2.sspro.p1.C.D.cond10\endcsname{0.074}
\expandafter\def\csname cnt@rev.desc.d2.sspro.p1.C.D.cond10\endcsname{0}
\expandafter\def\csname pct@rev.desc.d2.sspro.p1.C.D.cond10\endcsname{7.4}
\expandafter\def\csname ppt@rev.desc.d2.sspro.p1.C.D.cond10\endcsname{$+$7.4}
\expandafter\def\csname val@rev.desc.d2.sspro.p1.C.D.n\endcsname{48}
\expandafter\def\csname cnt@rev.desc.d2.sspro.p1.C.D.n\endcsname{48}
\expandafter\def\csname pct@rev.desc.d2.sspro.p1.C.D.n\endcsname{4800.0}
\expandafter\def\csname ppt@rev.desc.d2.sspro.p1.C.D.n\endcsname{$+$4800.0}
\expandafter\def\csname val@rev.desc.d2.sspro.p1.C.D.n-apps\endcsname{6}
\expandafter\def\csname cnt@rev.desc.d2.sspro.p1.C.D.n-apps\endcsname{6}
\expandafter\def\csname pct@rev.desc.d2.sspro.p1.C.D.n-apps\endcsname{600.0}
\expandafter\def\csname ppt@rev.desc.d2.sspro.p1.C.D.n-apps\endcsname{$+$600.0}
\expandafter\def\csname val@rev.desc.d2.sspro.p1.C.D.n01\endcsname{6}
\expandafter\def\csname cnt@rev.desc.d2.sspro.p1.C.D.n01\endcsname{6}
\expandafter\def\csname pct@rev.desc.d2.sspro.p1.C.D.n01\endcsname{600.0}
\expandafter\def\csname ppt@rev.desc.d2.sspro.p1.C.D.n01\endcsname{$+$600.0}
\expandafter\def\csname val@rev.desc.d2.sspro.p1.C.D.n10\endcsname{2}
\expandafter\def\csname cnt@rev.desc.d2.sspro.p1.C.D.n10\endcsname{2}
\expandafter\def\csname pct@rev.desc.d2.sspro.p1.C.D.n10\endcsname{200.0}
\expandafter\def\csname ppt@rev.desc.d2.sspro.p1.C.D.n10\endcsname{$+$200.0}
\expandafter\def\csname val@rev.desc.d2.sspro.p1.C.D.rate-any\endcsname{0.167}
\expandafter\def\csname cnt@rev.desc.d2.sspro.p1.C.D.rate-any\endcsname{0}
\expandafter\def\csname pct@rev.desc.d2.sspro.p1.C.D.rate-any\endcsname{16.7}
\expandafter\def\csname ppt@rev.desc.d2.sspro.p1.C.D.rate-any\endcsname{$+$16.7}
\expandafter\def\csname val@rev.desc.d2.sspro.p1.C.D.rate01\endcsname{0.125}
\expandafter\def\csname cnt@rev.desc.d2.sspro.p1.C.D.rate01\endcsname{0}
\expandafter\def\csname pct@rev.desc.d2.sspro.p1.C.D.rate01\endcsname{12.5}
\expandafter\def\csname ppt@rev.desc.d2.sspro.p1.C.D.rate01\endcsname{$+$12.5}
\expandafter\def\csname val@rev.desc.d2.sspro.p1.C.D.rate10\endcsname{0.042}
\expandafter\def\csname cnt@rev.desc.d2.sspro.p1.C.D.rate10\endcsname{0}
\expandafter\def\csname pct@rev.desc.d2.sspro.p1.C.D.rate10\endcsname{4.2}
\expandafter\def\csname ppt@rev.desc.d2.sspro.p1.C.D.rate10\endcsname{$+$4.2}
\expandafter\def\csname val@rev.desc.d2.sspro.p1.C.F.center-d1.acc-a\endcsname{0.708}
\expandafter\def\csname cnt@rev.desc.d2.sspro.p1.C.F.center-d1.acc-a\endcsname{1}
\expandafter\def\csname pct@rev.desc.d2.sspro.p1.C.F.center-d1.acc-a\endcsname{70.8}
\expandafter\def\csname ppt@rev.desc.d2.sspro.p1.C.F.center-d1.acc-a\endcsname{$+$70.8}
\expandafter\def\csname val@rev.desc.d2.sspro.p1.C.F.center-d1.acc-b\endcsname{0.812}
\expandafter\def\csname cnt@rev.desc.d2.sspro.p1.C.F.center-d1.acc-b\endcsname{1}
\expandafter\def\csname pct@rev.desc.d2.sspro.p1.C.F.center-d1.acc-b\endcsname{81.2}
\expandafter\def\csname ppt@rev.desc.d2.sspro.p1.C.F.center-d1.acc-b\endcsname{$+$81.2}
\expandafter\def\csname val@rev.desc.d2.sspro.p1.C.F.center-d1.cond01\endcsname{0.429}
\expandafter\def\csname cnt@rev.desc.d2.sspro.p1.C.F.center-d1.cond01\endcsname{0}
\expandafter\def\csname pct@rev.desc.d2.sspro.p1.C.F.center-d1.cond01\endcsname{42.9}
\expandafter\def\csname ppt@rev.desc.d2.sspro.p1.C.F.center-d1.cond01\endcsname{$+$42.9}
\expandafter\def\csname val@rev.desc.d2.sspro.p1.C.F.center-d1.cond10\endcsname{0.029}
\expandafter\def\csname cnt@rev.desc.d2.sspro.p1.C.F.center-d1.cond10\endcsname{0}
\expandafter\def\csname pct@rev.desc.d2.sspro.p1.C.F.center-d1.cond10\endcsname{2.9}
\expandafter\def\csname ppt@rev.desc.d2.sspro.p1.C.F.center-d1.cond10\endcsname{$+$2.9}
\expandafter\def\csname val@rev.desc.d2.sspro.p1.C.F.center-d1.n\endcsname{48}
\expandafter\def\csname cnt@rev.desc.d2.sspro.p1.C.F.center-d1.n\endcsname{48}
\expandafter\def\csname pct@rev.desc.d2.sspro.p1.C.F.center-d1.n\endcsname{4800.0}
\expandafter\def\csname ppt@rev.desc.d2.sspro.p1.C.F.center-d1.n\endcsname{$+$4800.0}
\expandafter\def\csname val@rev.desc.d2.sspro.p1.C.F.center-d1.n-apps\endcsname{6}
\expandafter\def\csname cnt@rev.desc.d2.sspro.p1.C.F.center-d1.n-apps\endcsname{6}
\expandafter\def\csname pct@rev.desc.d2.sspro.p1.C.F.center-d1.n-apps\endcsname{600.0}
\expandafter\def\csname ppt@rev.desc.d2.sspro.p1.C.F.center-d1.n-apps\endcsname{$+$600.0}
\expandafter\def\csname val@rev.desc.d2.sspro.p1.C.F.center-d1.n01\endcsname{6}
\expandafter\def\csname cnt@rev.desc.d2.sspro.p1.C.F.center-d1.n01\endcsname{6}
\expandafter\def\csname pct@rev.desc.d2.sspro.p1.C.F.center-d1.n01\endcsname{600.0}
\expandafter\def\csname ppt@rev.desc.d2.sspro.p1.C.F.center-d1.n01\endcsname{$+$600.0}
\expandafter\def\csname val@rev.desc.d2.sspro.p1.C.F.center-d1.n10\endcsname{1}
\expandafter\def\csname cnt@rev.desc.d2.sspro.p1.C.F.center-d1.n10\endcsname{1}
\expandafter\def\csname pct@rev.desc.d2.sspro.p1.C.F.center-d1.n10\endcsname{100.0}
\expandafter\def\csname ppt@rev.desc.d2.sspro.p1.C.F.center-d1.n10\endcsname{$+$100.0}
\expandafter\def\csname val@rev.desc.d2.sspro.p1.C.F.center-d1.rate-any\endcsname{0.146}
\expandafter\def\csname cnt@rev.desc.d2.sspro.p1.C.F.center-d1.rate-any\endcsname{0}
\expandafter\def\csname pct@rev.desc.d2.sspro.p1.C.F.center-d1.rate-any\endcsname{14.6}
\expandafter\def\csname ppt@rev.desc.d2.sspro.p1.C.F.center-d1.rate-any\endcsname{$+$14.6}
\expandafter\def\csname val@rev.desc.d2.sspro.p1.C.F.center-d1.rate01\endcsname{0.125}
\expandafter\def\csname cnt@rev.desc.d2.sspro.p1.C.F.center-d1.rate01\endcsname{0}
\expandafter\def\csname pct@rev.desc.d2.sspro.p1.C.F.center-d1.rate01\endcsname{12.5}
\expandafter\def\csname ppt@rev.desc.d2.sspro.p1.C.F.center-d1.rate01\endcsname{$+$12.5}
\expandafter\def\csname val@rev.desc.d2.sspro.p1.C.F.center-d1.rate10\endcsname{0.021}
\expandafter\def\csname cnt@rev.desc.d2.sspro.p1.C.F.center-d1.rate10\endcsname{0}
\expandafter\def\csname pct@rev.desc.d2.sspro.p1.C.F.center-d1.rate10\endcsname{2.1}
\expandafter\def\csname ppt@rev.desc.d2.sspro.p1.C.F.center-d1.rate10\endcsname{$+$2.1}
\expandafter\def\csname val@rev.desc.d2.sspro.p1.C.F.context-d1.acc-a\endcsname{0.729}
\expandafter\def\csname cnt@rev.desc.d2.sspro.p1.C.F.context-d1.acc-a\endcsname{1}
\expandafter\def\csname pct@rev.desc.d2.sspro.p1.C.F.context-d1.acc-a\endcsname{72.9}
\expandafter\def\csname ppt@rev.desc.d2.sspro.p1.C.F.context-d1.acc-a\endcsname{$+$72.9}
\expandafter\def\csname val@rev.desc.d2.sspro.p1.C.F.context-d1.acc-b\endcsname{0.750}
\expandafter\def\csname cnt@rev.desc.d2.sspro.p1.C.F.context-d1.acc-b\endcsname{1}
\expandafter\def\csname pct@rev.desc.d2.sspro.p1.C.F.context-d1.acc-b\endcsname{75.0}
\expandafter\def\csname ppt@rev.desc.d2.sspro.p1.C.F.context-d1.acc-b\endcsname{$+$75.0}
\expandafter\def\csname val@rev.desc.d2.sspro.p1.C.F.context-d1.cond01\endcsname{0.231}
\expandafter\def\csname cnt@rev.desc.d2.sspro.p1.C.F.context-d1.cond01\endcsname{0}
\expandafter\def\csname pct@rev.desc.d2.sspro.p1.C.F.context-d1.cond01\endcsname{23.1}
\expandafter\def\csname ppt@rev.desc.d2.sspro.p1.C.F.context-d1.cond01\endcsname{$+$23.1}
\expandafter\def\csname val@rev.desc.d2.sspro.p1.C.F.context-d1.cond10\endcsname{0.057}
\expandafter\def\csname cnt@rev.desc.d2.sspro.p1.C.F.context-d1.cond10\endcsname{0}
\expandafter\def\csname pct@rev.desc.d2.sspro.p1.C.F.context-d1.cond10\endcsname{5.7}
\expandafter\def\csname ppt@rev.desc.d2.sspro.p1.C.F.context-d1.cond10\endcsname{$+$5.7}
\expandafter\def\csname val@rev.desc.d2.sspro.p1.C.F.context-d1.n\endcsname{48}
\expandafter\def\csname cnt@rev.desc.d2.sspro.p1.C.F.context-d1.n\endcsname{48}
\expandafter\def\csname pct@rev.desc.d2.sspro.p1.C.F.context-d1.n\endcsname{4800.0}
\expandafter\def\csname ppt@rev.desc.d2.sspro.p1.C.F.context-d1.n\endcsname{$+$4800.0}
\expandafter\def\csname val@rev.desc.d2.sspro.p1.C.F.context-d1.n-apps\endcsname{6}
\expandafter\def\csname cnt@rev.desc.d2.sspro.p1.C.F.context-d1.n-apps\endcsname{6}
\expandafter\def\csname pct@rev.desc.d2.sspro.p1.C.F.context-d1.n-apps\endcsname{600.0}
\expandafter\def\csname ppt@rev.desc.d2.sspro.p1.C.F.context-d1.n-apps\endcsname{$+$600.0}
\expandafter\def\csname val@rev.desc.d2.sspro.p1.C.F.context-d1.n01\endcsname{3}
\expandafter\def\csname cnt@rev.desc.d2.sspro.p1.C.F.context-d1.n01\endcsname{3}
\expandafter\def\csname pct@rev.desc.d2.sspro.p1.C.F.context-d1.n01\endcsname{300.0}
\expandafter\def\csname ppt@rev.desc.d2.sspro.p1.C.F.context-d1.n01\endcsname{$+$300.0}
\expandafter\def\csname val@rev.desc.d2.sspro.p1.C.F.context-d1.n10\endcsname{2}
\expandafter\def\csname cnt@rev.desc.d2.sspro.p1.C.F.context-d1.n10\endcsname{2}
\expandafter\def\csname pct@rev.desc.d2.sspro.p1.C.F.context-d1.n10\endcsname{200.0}
\expandafter\def\csname ppt@rev.desc.d2.sspro.p1.C.F.context-d1.n10\endcsname{$+$200.0}
\expandafter\def\csname val@rev.desc.d2.sspro.p1.C.F.context-d1.rate-any\endcsname{0.104}
\expandafter\def\csname cnt@rev.desc.d2.sspro.p1.C.F.context-d1.rate-any\endcsname{0}
\expandafter\def\csname pct@rev.desc.d2.sspro.p1.C.F.context-d1.rate-any\endcsname{10.4}
\expandafter\def\csname ppt@rev.desc.d2.sspro.p1.C.F.context-d1.rate-any\endcsname{$+$10.4}
\expandafter\def\csname val@rev.desc.d2.sspro.p1.C.F.context-d1.rate01\endcsname{0.062}
\expandafter\def\csname cnt@rev.desc.d2.sspro.p1.C.F.context-d1.rate01\endcsname{0}
\expandafter\def\csname pct@rev.desc.d2.sspro.p1.C.F.context-d1.rate01\endcsname{6.2}
\expandafter\def\csname ppt@rev.desc.d2.sspro.p1.C.F.context-d1.rate01\endcsname{$+$6.2}
\expandafter\def\csname val@rev.desc.d2.sspro.p1.C.F.context-d1.rate10\endcsname{0.042}
\expandafter\def\csname cnt@rev.desc.d2.sspro.p1.C.F.context-d1.rate10\endcsname{0}
\expandafter\def\csname pct@rev.desc.d2.sspro.p1.C.F.context-d1.rate10\endcsname{4.2}
\expandafter\def\csname ppt@rev.desc.d2.sspro.p1.C.F.context-d1.rate10\endcsname{$+$4.2}
\expandafter\def\csname val@rev.desc.d2.sspro.p1.C.F.dual-d2.acc-a\endcsname{0.688}
\expandafter\def\csname cnt@rev.desc.d2.sspro.p1.C.F.dual-d2.acc-a\endcsname{1}
\expandafter\def\csname pct@rev.desc.d2.sspro.p1.C.F.dual-d2.acc-a\endcsname{68.8}
\expandafter\def\csname ppt@rev.desc.d2.sspro.p1.C.F.dual-d2.acc-a\endcsname{$+$68.8}
\expandafter\def\csname val@rev.desc.d2.sspro.p1.C.F.dual-d2.acc-b\endcsname{0.771}
\expandafter\def\csname cnt@rev.desc.d2.sspro.p1.C.F.dual-d2.acc-b\endcsname{1}
\expandafter\def\csname pct@rev.desc.d2.sspro.p1.C.F.dual-d2.acc-b\endcsname{77.1}
\expandafter\def\csname ppt@rev.desc.d2.sspro.p1.C.F.dual-d2.acc-b\endcsname{$+$77.1}
\expandafter\def\csname val@rev.desc.d2.sspro.p1.C.F.dual-d2.cond01\endcsname{0.400}
\expandafter\def\csname cnt@rev.desc.d2.sspro.p1.C.F.dual-d2.cond01\endcsname{0}
\expandafter\def\csname pct@rev.desc.d2.sspro.p1.C.F.dual-d2.cond01\endcsname{40.0}
\expandafter\def\csname ppt@rev.desc.d2.sspro.p1.C.F.dual-d2.cond01\endcsname{$+$40.0}
\expandafter\def\csname val@rev.desc.d2.sspro.p1.C.F.dual-d2.cond10\endcsname{0.061}
\expandafter\def\csname cnt@rev.desc.d2.sspro.p1.C.F.dual-d2.cond10\endcsname{0}
\expandafter\def\csname pct@rev.desc.d2.sspro.p1.C.F.dual-d2.cond10\endcsname{6.1}
\expandafter\def\csname ppt@rev.desc.d2.sspro.p1.C.F.dual-d2.cond10\endcsname{$+$6.1}
\expandafter\def\csname val@rev.desc.d2.sspro.p1.C.F.dual-d2.n\endcsname{48}
\expandafter\def\csname cnt@rev.desc.d2.sspro.p1.C.F.dual-d2.n\endcsname{48}
\expandafter\def\csname pct@rev.desc.d2.sspro.p1.C.F.dual-d2.n\endcsname{4800.0}
\expandafter\def\csname ppt@rev.desc.d2.sspro.p1.C.F.dual-d2.n\endcsname{$+$4800.0}
\expandafter\def\csname val@rev.desc.d2.sspro.p1.C.F.dual-d2.n-apps\endcsname{6}
\expandafter\def\csname cnt@rev.desc.d2.sspro.p1.C.F.dual-d2.n-apps\endcsname{6}
\expandafter\def\csname pct@rev.desc.d2.sspro.p1.C.F.dual-d2.n-apps\endcsname{600.0}
\expandafter\def\csname ppt@rev.desc.d2.sspro.p1.C.F.dual-d2.n-apps\endcsname{$+$600.0}
\expandafter\def\csname val@rev.desc.d2.sspro.p1.C.F.dual-d2.n01\endcsname{6}
\expandafter\def\csname cnt@rev.desc.d2.sspro.p1.C.F.dual-d2.n01\endcsname{6}
\expandafter\def\csname pct@rev.desc.d2.sspro.p1.C.F.dual-d2.n01\endcsname{600.0}
\expandafter\def\csname ppt@rev.desc.d2.sspro.p1.C.F.dual-d2.n01\endcsname{$+$600.0}
\expandafter\def\csname val@rev.desc.d2.sspro.p1.C.F.dual-d2.n10\endcsname{2}
\expandafter\def\csname cnt@rev.desc.d2.sspro.p1.C.F.dual-d2.n10\endcsname{2}
\expandafter\def\csname pct@rev.desc.d2.sspro.p1.C.F.dual-d2.n10\endcsname{200.0}
\expandafter\def\csname ppt@rev.desc.d2.sspro.p1.C.F.dual-d2.n10\endcsname{$+$200.0}
\expandafter\def\csname val@rev.desc.d2.sspro.p1.C.F.dual-d2.rate-any\endcsname{0.167}
\expandafter\def\csname cnt@rev.desc.d2.sspro.p1.C.F.dual-d2.rate-any\endcsname{0}
\expandafter\def\csname pct@rev.desc.d2.sspro.p1.C.F.dual-d2.rate-any\endcsname{16.7}
\expandafter\def\csname ppt@rev.desc.d2.sspro.p1.C.F.dual-d2.rate-any\endcsname{$+$16.7}
\expandafter\def\csname val@rev.desc.d2.sspro.p1.C.F.dual-d2.rate01\endcsname{0.125}
\expandafter\def\csname cnt@rev.desc.d2.sspro.p1.C.F.dual-d2.rate01\endcsname{0}
\expandafter\def\csname pct@rev.desc.d2.sspro.p1.C.F.dual-d2.rate01\endcsname{12.5}
\expandafter\def\csname ppt@rev.desc.d2.sspro.p1.C.F.dual-d2.rate01\endcsname{$+$12.5}
\expandafter\def\csname val@rev.desc.d2.sspro.p1.C.F.dual-d2.rate10\endcsname{0.042}
\expandafter\def\csname cnt@rev.desc.d2.sspro.p1.C.F.dual-d2.rate10\endcsname{0}
\expandafter\def\csname pct@rev.desc.d2.sspro.p1.C.F.dual-d2.rate10\endcsname{4.2}
\expandafter\def\csname ppt@rev.desc.d2.sspro.p1.C.F.dual-d2.rate10\endcsname{$+$4.2}
\expandafter\def\csname val@rev.desc.d2.sspro.p1.C.F.grid2-d3.acc-a\endcsname{0.771}
\expandafter\def\csname cnt@rev.desc.d2.sspro.p1.C.F.grid2-d3.acc-a\endcsname{1}
\expandafter\def\csname pct@rev.desc.d2.sspro.p1.C.F.grid2-d3.acc-a\endcsname{77.1}
\expandafter\def\csname ppt@rev.desc.d2.sspro.p1.C.F.grid2-d3.acc-a\endcsname{$+$77.1}
\expandafter\def\csname val@rev.desc.d2.sspro.p1.C.F.grid2-d3.acc-b\endcsname{0.792}
\expandafter\def\csname cnt@rev.desc.d2.sspro.p1.C.F.grid2-d3.acc-b\endcsname{1}
\expandafter\def\csname pct@rev.desc.d2.sspro.p1.C.F.grid2-d3.acc-b\endcsname{79.2}
\expandafter\def\csname ppt@rev.desc.d2.sspro.p1.C.F.grid2-d3.acc-b\endcsname{$+$79.2}
\expandafter\def\csname val@rev.desc.d2.sspro.p1.C.F.grid2-d3.cond01\endcsname{0.273}
\expandafter\def\csname cnt@rev.desc.d2.sspro.p1.C.F.grid2-d3.cond01\endcsname{0}
\expandafter\def\csname pct@rev.desc.d2.sspro.p1.C.F.grid2-d3.cond01\endcsname{27.3}
\expandafter\def\csname ppt@rev.desc.d2.sspro.p1.C.F.grid2-d3.cond01\endcsname{$+$27.3}
\expandafter\def\csname val@rev.desc.d2.sspro.p1.C.F.grid2-d3.cond10\endcsname{0.054}
\expandafter\def\csname cnt@rev.desc.d2.sspro.p1.C.F.grid2-d3.cond10\endcsname{0}
\expandafter\def\csname pct@rev.desc.d2.sspro.p1.C.F.grid2-d3.cond10\endcsname{5.4}
\expandafter\def\csname ppt@rev.desc.d2.sspro.p1.C.F.grid2-d3.cond10\endcsname{$+$5.4}
\expandafter\def\csname val@rev.desc.d2.sspro.p1.C.F.grid2-d3.n\endcsname{48}
\expandafter\def\csname cnt@rev.desc.d2.sspro.p1.C.F.grid2-d3.n\endcsname{48}
\expandafter\def\csname pct@rev.desc.d2.sspro.p1.C.F.grid2-d3.n\endcsname{4800.0}
\expandafter\def\csname ppt@rev.desc.d2.sspro.p1.C.F.grid2-d3.n\endcsname{$+$4800.0}
\expandafter\def\csname val@rev.desc.d2.sspro.p1.C.F.grid2-d3.n-apps\endcsname{6}
\expandafter\def\csname cnt@rev.desc.d2.sspro.p1.C.F.grid2-d3.n-apps\endcsname{6}
\expandafter\def\csname pct@rev.desc.d2.sspro.p1.C.F.grid2-d3.n-apps\endcsname{600.0}
\expandafter\def\csname ppt@rev.desc.d2.sspro.p1.C.F.grid2-d3.n-apps\endcsname{$+$600.0}
\expandafter\def\csname val@rev.desc.d2.sspro.p1.C.F.grid2-d3.n01\endcsname{3}
\expandafter\def\csname cnt@rev.desc.d2.sspro.p1.C.F.grid2-d3.n01\endcsname{3}
\expandafter\def\csname pct@rev.desc.d2.sspro.p1.C.F.grid2-d3.n01\endcsname{300.0}
\expandafter\def\csname ppt@rev.desc.d2.sspro.p1.C.F.grid2-d3.n01\endcsname{$+$300.0}
\expandafter\def\csname val@rev.desc.d2.sspro.p1.C.F.grid2-d3.n10\endcsname{2}
\expandafter\def\csname cnt@rev.desc.d2.sspro.p1.C.F.grid2-d3.n10\endcsname{2}
\expandafter\def\csname pct@rev.desc.d2.sspro.p1.C.F.grid2-d3.n10\endcsname{200.0}
\expandafter\def\csname ppt@rev.desc.d2.sspro.p1.C.F.grid2-d3.n10\endcsname{$+$200.0}
\expandafter\def\csname val@rev.desc.d2.sspro.p1.C.F.grid2-d3.rate-any\endcsname{0.104}
\expandafter\def\csname cnt@rev.desc.d2.sspro.p1.C.F.grid2-d3.rate-any\endcsname{0}
\expandafter\def\csname pct@rev.desc.d2.sspro.p1.C.F.grid2-d3.rate-any\endcsname{10.4}
\expandafter\def\csname ppt@rev.desc.d2.sspro.p1.C.F.grid2-d3.rate-any\endcsname{$+$10.4}
\expandafter\def\csname val@rev.desc.d2.sspro.p1.C.F.grid2-d3.rate01\endcsname{0.062}
\expandafter\def\csname cnt@rev.desc.d2.sspro.p1.C.F.grid2-d3.rate01\endcsname{0}
\expandafter\def\csname pct@rev.desc.d2.sspro.p1.C.F.grid2-d3.rate01\endcsname{6.2}
\expandafter\def\csname ppt@rev.desc.d2.sspro.p1.C.F.grid2-d3.rate01\endcsname{$+$6.2}
\expandafter\def\csname val@rev.desc.d2.sspro.p1.C.F.grid2-d3.rate10\endcsname{0.042}
\expandafter\def\csname cnt@rev.desc.d2.sspro.p1.C.F.grid2-d3.rate10\endcsname{0}
\expandafter\def\csname pct@rev.desc.d2.sspro.p1.C.F.grid2-d3.rate10\endcsname{4.2}
\expandafter\def\csname ppt@rev.desc.d2.sspro.p1.C.F.grid2-d3.rate10\endcsname{$+$4.2}
\expandafter\def\csname val@rev.desc.d2.sspro.p1.C.F.zoomclick-d3.acc-a\endcsname{0.812}
\expandafter\def\csname cnt@rev.desc.d2.sspro.p1.C.F.zoomclick-d3.acc-a\endcsname{1}
\expandafter\def\csname pct@rev.desc.d2.sspro.p1.C.F.zoomclick-d3.acc-a\endcsname{81.2}
\expandafter\def\csname ppt@rev.desc.d2.sspro.p1.C.F.zoomclick-d3.acc-a\endcsname{$+$81.2}
\expandafter\def\csname val@rev.desc.d2.sspro.p1.C.F.zoomclick-d3.acc-b\endcsname{0.854}
\expandafter\def\csname cnt@rev.desc.d2.sspro.p1.C.F.zoomclick-d3.acc-b\endcsname{1}
\expandafter\def\csname pct@rev.desc.d2.sspro.p1.C.F.zoomclick-d3.acc-b\endcsname{85.4}
\expandafter\def\csname ppt@rev.desc.d2.sspro.p1.C.F.zoomclick-d3.acc-b\endcsname{$+$85.4}
\expandafter\def\csname val@rev.desc.d2.sspro.p1.C.F.zoomclick-d3.cond01\endcsname{0.222}
\expandafter\def\csname cnt@rev.desc.d2.sspro.p1.C.F.zoomclick-d3.cond01\endcsname{0}
\expandafter\def\csname pct@rev.desc.d2.sspro.p1.C.F.zoomclick-d3.cond01\endcsname{22.2}
\expandafter\def\csname ppt@rev.desc.d2.sspro.p1.C.F.zoomclick-d3.cond01\endcsname{$+$22.2}
\expandafter\def\csname val@rev.desc.d2.sspro.p1.C.F.zoomclick-d3.cond10\endcsname{0.000}
\expandafter\def\csname cnt@rev.desc.d2.sspro.p1.C.F.zoomclick-d3.cond10\endcsname{0}
\expandafter\def\csname pct@rev.desc.d2.sspro.p1.C.F.zoomclick-d3.cond10\endcsname{0.0}
\expandafter\def\csname ppt@rev.desc.d2.sspro.p1.C.F.zoomclick-d3.cond10\endcsname{0.0}
\expandafter\def\csname val@rev.desc.d2.sspro.p1.C.F.zoomclick-d3.n\endcsname{48}
\expandafter\def\csname cnt@rev.desc.d2.sspro.p1.C.F.zoomclick-d3.n\endcsname{48}
\expandafter\def\csname pct@rev.desc.d2.sspro.p1.C.F.zoomclick-d3.n\endcsname{4800.0}
\expandafter\def\csname ppt@rev.desc.d2.sspro.p1.C.F.zoomclick-d3.n\endcsname{$+$4800.0}
\expandafter\def\csname val@rev.desc.d2.sspro.p1.C.F.zoomclick-d3.n-apps\endcsname{6}
\expandafter\def\csname cnt@rev.desc.d2.sspro.p1.C.F.zoomclick-d3.n-apps\endcsname{6}
\expandafter\def\csname pct@rev.desc.d2.sspro.p1.C.F.zoomclick-d3.n-apps\endcsname{600.0}
\expandafter\def\csname ppt@rev.desc.d2.sspro.p1.C.F.zoomclick-d3.n-apps\endcsname{$+$600.0}
\expandafter\def\csname val@rev.desc.d2.sspro.p1.C.F.zoomclick-d3.n01\endcsname{2}
\expandafter\def\csname cnt@rev.desc.d2.sspro.p1.C.F.zoomclick-d3.n01\endcsname{2}
\expandafter\def\csname pct@rev.desc.d2.sspro.p1.C.F.zoomclick-d3.n01\endcsname{200.0}
\expandafter\def\csname ppt@rev.desc.d2.sspro.p1.C.F.zoomclick-d3.n01\endcsname{$+$200.0}
\expandafter\def\csname val@rev.desc.d2.sspro.p1.C.F.zoomclick-d3.n10\endcsname{0}
\expandafter\def\csname cnt@rev.desc.d2.sspro.p1.C.F.zoomclick-d3.n10\endcsname{0}
\expandafter\def\csname pct@rev.desc.d2.sspro.p1.C.F.zoomclick-d3.n10\endcsname{0.0}
\expandafter\def\csname ppt@rev.desc.d2.sspro.p1.C.F.zoomclick-d3.n10\endcsname{0.0}
\expandafter\def\csname val@rev.desc.d2.sspro.p1.C.F.zoomclick-d3.rate-any\endcsname{0.042}
\expandafter\def\csname cnt@rev.desc.d2.sspro.p1.C.F.zoomclick-d3.rate-any\endcsname{0}
\expandafter\def\csname pct@rev.desc.d2.sspro.p1.C.F.zoomclick-d3.rate-any\endcsname{4.2}
\expandafter\def\csname ppt@rev.desc.d2.sspro.p1.C.F.zoomclick-d3.rate-any\endcsname{$+$4.2}
\expandafter\def\csname val@rev.desc.d2.sspro.p1.C.F.zoomclick-d3.rate01\endcsname{0.042}
\expandafter\def\csname cnt@rev.desc.d2.sspro.p1.C.F.zoomclick-d3.rate01\endcsname{0}
\expandafter\def\csname pct@rev.desc.d2.sspro.p1.C.F.zoomclick-d3.rate01\endcsname{4.2}
\expandafter\def\csname ppt@rev.desc.d2.sspro.p1.C.F.zoomclick-d3.rate01\endcsname{$+$4.2}
\expandafter\def\csname val@rev.desc.d2.sspro.p1.C.F.zoomclick-d3.rate10\endcsname{0.000}
\expandafter\def\csname cnt@rev.desc.d2.sspro.p1.C.F.zoomclick-d3.rate10\endcsname{0}
\expandafter\def\csname pct@rev.desc.d2.sspro.p1.C.F.zoomclick-d3.rate10\endcsname{0.0}
\expandafter\def\csname ppt@rev.desc.d2.sspro.p1.C.F.zoomclick-d3.rate10\endcsname{0.0}
\expandafter\def\csname val@rev.desc.d2.sspro.p1.T.D.acc-a\endcsname{0.542}
\expandafter\def\csname cnt@rev.desc.d2.sspro.p1.T.D.acc-a\endcsname{1}
\expandafter\def\csname pct@rev.desc.d2.sspro.p1.T.D.acc-a\endcsname{54.2}
\expandafter\def\csname ppt@rev.desc.d2.sspro.p1.T.D.acc-a\endcsname{$+$54.2}
\expandafter\def\csname val@rev.desc.d2.sspro.p1.T.D.acc-b\endcsname{0.740}
\expandafter\def\csname cnt@rev.desc.d2.sspro.p1.T.D.acc-b\endcsname{1}
\expandafter\def\csname pct@rev.desc.d2.sspro.p1.T.D.acc-b\endcsname{74.0}
\expandafter\def\csname ppt@rev.desc.d2.sspro.p1.T.D.acc-b\endcsname{$+$74.0}
\expandafter\def\csname val@rev.desc.d2.sspro.p1.T.D.cond01\endcsname{0.500}
\expandafter\def\csname cnt@rev.desc.d2.sspro.p1.T.D.cond01\endcsname{0}
\expandafter\def\csname pct@rev.desc.d2.sspro.p1.T.D.cond01\endcsname{50.0}
\expandafter\def\csname ppt@rev.desc.d2.sspro.p1.T.D.cond01\endcsname{$+$50.0}
\expandafter\def\csname val@rev.desc.d2.sspro.p1.T.D.cond10\endcsname{0.058}
\expandafter\def\csname cnt@rev.desc.d2.sspro.p1.T.D.cond10\endcsname{0}
\expandafter\def\csname pct@rev.desc.d2.sspro.p1.T.D.cond10\endcsname{5.8}
\expandafter\def\csname ppt@rev.desc.d2.sspro.p1.T.D.cond10\endcsname{$+$5.8}
\expandafter\def\csname val@rev.desc.d2.sspro.p1.T.D.n\endcsname{96}
\expandafter\def\csname cnt@rev.desc.d2.sspro.p1.T.D.n\endcsname{96}
\expandafter\def\csname pct@rev.desc.d2.sspro.p1.T.D.n\endcsname{9600.0}
\expandafter\def\csname ppt@rev.desc.d2.sspro.p1.T.D.n\endcsname{$+$9600.0}
\expandafter\def\csname val@rev.desc.d2.sspro.p1.T.D.n-apps\endcsname{12}
\expandafter\def\csname cnt@rev.desc.d2.sspro.p1.T.D.n-apps\endcsname{12}
\expandafter\def\csname pct@rev.desc.d2.sspro.p1.T.D.n-apps\endcsname{1200.0}
\expandafter\def\csname ppt@rev.desc.d2.sspro.p1.T.D.n-apps\endcsname{$+$1200.0}
\expandafter\def\csname val@rev.desc.d2.sspro.p1.T.D.n01\endcsname{22}
\expandafter\def\csname cnt@rev.desc.d2.sspro.p1.T.D.n01\endcsname{22}
\expandafter\def\csname pct@rev.desc.d2.sspro.p1.T.D.n01\endcsname{2200.0}
\expandafter\def\csname ppt@rev.desc.d2.sspro.p1.T.D.n01\endcsname{$+$2200.0}
\expandafter\def\csname val@rev.desc.d2.sspro.p1.T.D.n10\endcsname{3}
\expandafter\def\csname cnt@rev.desc.d2.sspro.p1.T.D.n10\endcsname{3}
\expandafter\def\csname pct@rev.desc.d2.sspro.p1.T.D.n10\endcsname{300.0}
\expandafter\def\csname ppt@rev.desc.d2.sspro.p1.T.D.n10\endcsname{$+$300.0}
\expandafter\def\csname val@rev.desc.d2.sspro.p1.T.D.rate-any\endcsname{0.260}
\expandafter\def\csname cnt@rev.desc.d2.sspro.p1.T.D.rate-any\endcsname{0}
\expandafter\def\csname pct@rev.desc.d2.sspro.p1.T.D.rate-any\endcsname{26.0}
\expandafter\def\csname ppt@rev.desc.d2.sspro.p1.T.D.rate-any\endcsname{$+$26.0}
\expandafter\def\csname val@rev.desc.d2.sspro.p1.T.D.rate01\endcsname{0.229}
\expandafter\def\csname cnt@rev.desc.d2.sspro.p1.T.D.rate01\endcsname{0}
\expandafter\def\csname pct@rev.desc.d2.sspro.p1.T.D.rate01\endcsname{22.9}
\expandafter\def\csname ppt@rev.desc.d2.sspro.p1.T.D.rate01\endcsname{$+$22.9}
\expandafter\def\csname val@rev.desc.d2.sspro.p1.T.D.rate10\endcsname{0.031}
\expandafter\def\csname cnt@rev.desc.d2.sspro.p1.T.D.rate10\endcsname{0}
\expandafter\def\csname pct@rev.desc.d2.sspro.p1.T.D.rate10\endcsname{3.1}
\expandafter\def\csname ppt@rev.desc.d2.sspro.p1.T.D.rate10\endcsname{$+$3.1}
\expandafter\def\csname val@rev.desc.d2.sspro.p2.C.D.acc-a\endcsname{0.646}
\expandafter\def\csname cnt@rev.desc.d2.sspro.p2.C.D.acc-a\endcsname{1}
\expandafter\def\csname pct@rev.desc.d2.sspro.p2.C.D.acc-a\endcsname{64.6}
\expandafter\def\csname ppt@rev.desc.d2.sspro.p2.C.D.acc-a\endcsname{$+$64.6}
\expandafter\def\csname val@rev.desc.d2.sspro.p2.C.D.acc-b\endcsname{0.729}
\expandafter\def\csname cnt@rev.desc.d2.sspro.p2.C.D.acc-b\endcsname{1}
\expandafter\def\csname pct@rev.desc.d2.sspro.p2.C.D.acc-b\endcsname{72.9}
\expandafter\def\csname ppt@rev.desc.d2.sspro.p2.C.D.acc-b\endcsname{$+$72.9}
\expandafter\def\csname val@rev.desc.d2.sspro.p2.C.D.cond01\endcsname{0.294}
\expandafter\def\csname cnt@rev.desc.d2.sspro.p2.C.D.cond01\endcsname{0}
\expandafter\def\csname pct@rev.desc.d2.sspro.p2.C.D.cond01\endcsname{29.4}
\expandafter\def\csname ppt@rev.desc.d2.sspro.p2.C.D.cond01\endcsname{$+$29.4}
\expandafter\def\csname val@rev.desc.d2.sspro.p2.C.D.cond10\endcsname{0.032}
\expandafter\def\csname cnt@rev.desc.d2.sspro.p2.C.D.cond10\endcsname{0}
\expandafter\def\csname pct@rev.desc.d2.sspro.p2.C.D.cond10\endcsname{3.2}
\expandafter\def\csname ppt@rev.desc.d2.sspro.p2.C.D.cond10\endcsname{$+$3.2}
\expandafter\def\csname val@rev.desc.d2.sspro.p2.C.D.n\endcsname{48}
\expandafter\def\csname cnt@rev.desc.d2.sspro.p2.C.D.n\endcsname{48}
\expandafter\def\csname pct@rev.desc.d2.sspro.p2.C.D.n\endcsname{4800.0}
\expandafter\def\csname ppt@rev.desc.d2.sspro.p2.C.D.n\endcsname{$+$4800.0}
\expandafter\def\csname val@rev.desc.d2.sspro.p2.C.D.n-apps\endcsname{6}
\expandafter\def\csname cnt@rev.desc.d2.sspro.p2.C.D.n-apps\endcsname{6}
\expandafter\def\csname pct@rev.desc.d2.sspro.p2.C.D.n-apps\endcsname{600.0}
\expandafter\def\csname ppt@rev.desc.d2.sspro.p2.C.D.n-apps\endcsname{$+$600.0}
\expandafter\def\csname val@rev.desc.d2.sspro.p2.C.D.n01\endcsname{5}
\expandafter\def\csname cnt@rev.desc.d2.sspro.p2.C.D.n01\endcsname{5}
\expandafter\def\csname pct@rev.desc.d2.sspro.p2.C.D.n01\endcsname{500.0}
\expandafter\def\csname ppt@rev.desc.d2.sspro.p2.C.D.n01\endcsname{$+$500.0}
\expandafter\def\csname val@rev.desc.d2.sspro.p2.C.D.n10\endcsname{1}
\expandafter\def\csname cnt@rev.desc.d2.sspro.p2.C.D.n10\endcsname{1}
\expandafter\def\csname pct@rev.desc.d2.sspro.p2.C.D.n10\endcsname{100.0}
\expandafter\def\csname ppt@rev.desc.d2.sspro.p2.C.D.n10\endcsname{$+$100.0}
\expandafter\def\csname val@rev.desc.d2.sspro.p2.C.D.rate-any\endcsname{0.125}
\expandafter\def\csname cnt@rev.desc.d2.sspro.p2.C.D.rate-any\endcsname{0}
\expandafter\def\csname pct@rev.desc.d2.sspro.p2.C.D.rate-any\endcsname{12.5}
\expandafter\def\csname ppt@rev.desc.d2.sspro.p2.C.D.rate-any\endcsname{$+$12.5}
\expandafter\def\csname val@rev.desc.d2.sspro.p2.C.D.rate01\endcsname{0.104}
\expandafter\def\csname cnt@rev.desc.d2.sspro.p2.C.D.rate01\endcsname{0}
\expandafter\def\csname pct@rev.desc.d2.sspro.p2.C.D.rate01\endcsname{10.4}
\expandafter\def\csname ppt@rev.desc.d2.sspro.p2.C.D.rate01\endcsname{$+$10.4}
\expandafter\def\csname val@rev.desc.d2.sspro.p2.C.D.rate10\endcsname{0.021}
\expandafter\def\csname cnt@rev.desc.d2.sspro.p2.C.D.rate10\endcsname{0}
\expandafter\def\csname pct@rev.desc.d2.sspro.p2.C.D.rate10\endcsname{2.1}
\expandafter\def\csname ppt@rev.desc.d2.sspro.p2.C.D.rate10\endcsname{$+$2.1}
\expandafter\def\csname val@rev.desc.d2.sspro.p2.C.F.center-d1.acc-a\endcsname{0.812}
\expandafter\def\csname cnt@rev.desc.d2.sspro.p2.C.F.center-d1.acc-a\endcsname{1}
\expandafter\def\csname pct@rev.desc.d2.sspro.p2.C.F.center-d1.acc-a\endcsname{81.2}
\expandafter\def\csname ppt@rev.desc.d2.sspro.p2.C.F.center-d1.acc-a\endcsname{$+$81.2}
\expandafter\def\csname val@rev.desc.d2.sspro.p2.C.F.center-d1.acc-b\endcsname{0.833}
\expandafter\def\csname cnt@rev.desc.d2.sspro.p2.C.F.center-d1.acc-b\endcsname{1}
\expandafter\def\csname pct@rev.desc.d2.sspro.p2.C.F.center-d1.acc-b\endcsname{83.3}
\expandafter\def\csname ppt@rev.desc.d2.sspro.p2.C.F.center-d1.acc-b\endcsname{$+$83.3}
\expandafter\def\csname val@rev.desc.d2.sspro.p2.C.F.center-d1.cond01\endcsname{0.111}
\expandafter\def\csname cnt@rev.desc.d2.sspro.p2.C.F.center-d1.cond01\endcsname{0}
\expandafter\def\csname pct@rev.desc.d2.sspro.p2.C.F.center-d1.cond01\endcsname{11.1}
\expandafter\def\csname ppt@rev.desc.d2.sspro.p2.C.F.center-d1.cond01\endcsname{$+$11.1}
\expandafter\def\csname val@rev.desc.d2.sspro.p2.C.F.center-d1.cond10\endcsname{0.000}
\expandafter\def\csname cnt@rev.desc.d2.sspro.p2.C.F.center-d1.cond10\endcsname{0}
\expandafter\def\csname pct@rev.desc.d2.sspro.p2.C.F.center-d1.cond10\endcsname{0.0}
\expandafter\def\csname ppt@rev.desc.d2.sspro.p2.C.F.center-d1.cond10\endcsname{0.0}
\expandafter\def\csname val@rev.desc.d2.sspro.p2.C.F.center-d1.n\endcsname{48}
\expandafter\def\csname cnt@rev.desc.d2.sspro.p2.C.F.center-d1.n\endcsname{48}
\expandafter\def\csname pct@rev.desc.d2.sspro.p2.C.F.center-d1.n\endcsname{4800.0}
\expandafter\def\csname ppt@rev.desc.d2.sspro.p2.C.F.center-d1.n\endcsname{$+$4800.0}
\expandafter\def\csname val@rev.desc.d2.sspro.p2.C.F.center-d1.n-apps\endcsname{6}
\expandafter\def\csname cnt@rev.desc.d2.sspro.p2.C.F.center-d1.n-apps\endcsname{6}
\expandafter\def\csname pct@rev.desc.d2.sspro.p2.C.F.center-d1.n-apps\endcsname{600.0}
\expandafter\def\csname ppt@rev.desc.d2.sspro.p2.C.F.center-d1.n-apps\endcsname{$+$600.0}
\expandafter\def\csname val@rev.desc.d2.sspro.p2.C.F.center-d1.n01\endcsname{1}
\expandafter\def\csname cnt@rev.desc.d2.sspro.p2.C.F.center-d1.n01\endcsname{1}
\expandafter\def\csname pct@rev.desc.d2.sspro.p2.C.F.center-d1.n01\endcsname{100.0}
\expandafter\def\csname ppt@rev.desc.d2.sspro.p2.C.F.center-d1.n01\endcsname{$+$100.0}
\expandafter\def\csname val@rev.desc.d2.sspro.p2.C.F.center-d1.n10\endcsname{0}
\expandafter\def\csname cnt@rev.desc.d2.sspro.p2.C.F.center-d1.n10\endcsname{0}
\expandafter\def\csname pct@rev.desc.d2.sspro.p2.C.F.center-d1.n10\endcsname{0.0}
\expandafter\def\csname ppt@rev.desc.d2.sspro.p2.C.F.center-d1.n10\endcsname{0.0}
\expandafter\def\csname val@rev.desc.d2.sspro.p2.C.F.center-d1.rate-any\endcsname{0.021}
\expandafter\def\csname cnt@rev.desc.d2.sspro.p2.C.F.center-d1.rate-any\endcsname{0}
\expandafter\def\csname pct@rev.desc.d2.sspro.p2.C.F.center-d1.rate-any\endcsname{2.1}
\expandafter\def\csname ppt@rev.desc.d2.sspro.p2.C.F.center-d1.rate-any\endcsname{$+$2.1}
\expandafter\def\csname val@rev.desc.d2.sspro.p2.C.F.center-d1.rate01\endcsname{0.021}
\expandafter\def\csname cnt@rev.desc.d2.sspro.p2.C.F.center-d1.rate01\endcsname{0}
\expandafter\def\csname pct@rev.desc.d2.sspro.p2.C.F.center-d1.rate01\endcsname{2.1}
\expandafter\def\csname ppt@rev.desc.d2.sspro.p2.C.F.center-d1.rate01\endcsname{$+$2.1}
\expandafter\def\csname val@rev.desc.d2.sspro.p2.C.F.center-d1.rate10\endcsname{0.000}
\expandafter\def\csname cnt@rev.desc.d2.sspro.p2.C.F.center-d1.rate10\endcsname{0}
\expandafter\def\csname pct@rev.desc.d2.sspro.p2.C.F.center-d1.rate10\endcsname{0.0}
\expandafter\def\csname ppt@rev.desc.d2.sspro.p2.C.F.center-d1.rate10\endcsname{0.0}
\expandafter\def\csname val@rev.desc.d2.sspro.p2.C.F.context-d1.acc-a\endcsname{0.750}
\expandafter\def\csname cnt@rev.desc.d2.sspro.p2.C.F.context-d1.acc-a\endcsname{1}
\expandafter\def\csname pct@rev.desc.d2.sspro.p2.C.F.context-d1.acc-a\endcsname{75.0}
\expandafter\def\csname ppt@rev.desc.d2.sspro.p2.C.F.context-d1.acc-a\endcsname{$+$75.0}
\expandafter\def\csname val@rev.desc.d2.sspro.p2.C.F.context-d1.acc-b\endcsname{0.833}
\expandafter\def\csname cnt@rev.desc.d2.sspro.p2.C.F.context-d1.acc-b\endcsname{1}
\expandafter\def\csname pct@rev.desc.d2.sspro.p2.C.F.context-d1.acc-b\endcsname{83.3}
\expandafter\def\csname ppt@rev.desc.d2.sspro.p2.C.F.context-d1.acc-b\endcsname{$+$83.3}
\expandafter\def\csname val@rev.desc.d2.sspro.p2.C.F.context-d1.cond01\endcsname{0.333}
\expandafter\def\csname cnt@rev.desc.d2.sspro.p2.C.F.context-d1.cond01\endcsname{0}
\expandafter\def\csname pct@rev.desc.d2.sspro.p2.C.F.context-d1.cond01\endcsname{33.3}
\expandafter\def\csname ppt@rev.desc.d2.sspro.p2.C.F.context-d1.cond01\endcsname{$+$33.3}
\expandafter\def\csname val@rev.desc.d2.sspro.p2.C.F.context-d1.cond10\endcsname{0.000}
\expandafter\def\csname cnt@rev.desc.d2.sspro.p2.C.F.context-d1.cond10\endcsname{0}
\expandafter\def\csname pct@rev.desc.d2.sspro.p2.C.F.context-d1.cond10\endcsname{0.0}
\expandafter\def\csname ppt@rev.desc.d2.sspro.p2.C.F.context-d1.cond10\endcsname{0.0}
\expandafter\def\csname val@rev.desc.d2.sspro.p2.C.F.context-d1.n\endcsname{48}
\expandafter\def\csname cnt@rev.desc.d2.sspro.p2.C.F.context-d1.n\endcsname{48}
\expandafter\def\csname pct@rev.desc.d2.sspro.p2.C.F.context-d1.n\endcsname{4800.0}
\expandafter\def\csname ppt@rev.desc.d2.sspro.p2.C.F.context-d1.n\endcsname{$+$4800.0}
\expandafter\def\csname val@rev.desc.d2.sspro.p2.C.F.context-d1.n-apps\endcsname{6}
\expandafter\def\csname cnt@rev.desc.d2.sspro.p2.C.F.context-d1.n-apps\endcsname{6}
\expandafter\def\csname pct@rev.desc.d2.sspro.p2.C.F.context-d1.n-apps\endcsname{600.0}
\expandafter\def\csname ppt@rev.desc.d2.sspro.p2.C.F.context-d1.n-apps\endcsname{$+$600.0}
\expandafter\def\csname val@rev.desc.d2.sspro.p2.C.F.context-d1.n01\endcsname{4}
\expandafter\def\csname cnt@rev.desc.d2.sspro.p2.C.F.context-d1.n01\endcsname{4}
\expandafter\def\csname pct@rev.desc.d2.sspro.p2.C.F.context-d1.n01\endcsname{400.0}
\expandafter\def\csname ppt@rev.desc.d2.sspro.p2.C.F.context-d1.n01\endcsname{$+$400.0}
\expandafter\def\csname val@rev.desc.d2.sspro.p2.C.F.context-d1.n10\endcsname{0}
\expandafter\def\csname cnt@rev.desc.d2.sspro.p2.C.F.context-d1.n10\endcsname{0}
\expandafter\def\csname pct@rev.desc.d2.sspro.p2.C.F.context-d1.n10\endcsname{0.0}
\expandafter\def\csname ppt@rev.desc.d2.sspro.p2.C.F.context-d1.n10\endcsname{0.0}
\expandafter\def\csname val@rev.desc.d2.sspro.p2.C.F.context-d1.rate-any\endcsname{0.083}
\expandafter\def\csname cnt@rev.desc.d2.sspro.p2.C.F.context-d1.rate-any\endcsname{0}
\expandafter\def\csname pct@rev.desc.d2.sspro.p2.C.F.context-d1.rate-any\endcsname{8.3}
\expandafter\def\csname ppt@rev.desc.d2.sspro.p2.C.F.context-d1.rate-any\endcsname{$+$8.3}
\expandafter\def\csname val@rev.desc.d2.sspro.p2.C.F.context-d1.rate01\endcsname{0.083}
\expandafter\def\csname cnt@rev.desc.d2.sspro.p2.C.F.context-d1.rate01\endcsname{0}
\expandafter\def\csname pct@rev.desc.d2.sspro.p2.C.F.context-d1.rate01\endcsname{8.3}
\expandafter\def\csname ppt@rev.desc.d2.sspro.p2.C.F.context-d1.rate01\endcsname{$+$8.3}
\expandafter\def\csname val@rev.desc.d2.sspro.p2.C.F.context-d1.rate10\endcsname{0.000}
\expandafter\def\csname cnt@rev.desc.d2.sspro.p2.C.F.context-d1.rate10\endcsname{0}
\expandafter\def\csname pct@rev.desc.d2.sspro.p2.C.F.context-d1.rate10\endcsname{0.0}
\expandafter\def\csname ppt@rev.desc.d2.sspro.p2.C.F.context-d1.rate10\endcsname{0.0}
\expandafter\def\csname val@rev.desc.d2.sspro.p2.C.F.dual-d2.acc-a\endcsname{0.771}
\expandafter\def\csname cnt@rev.desc.d2.sspro.p2.C.F.dual-d2.acc-a\endcsname{1}
\expandafter\def\csname pct@rev.desc.d2.sspro.p2.C.F.dual-d2.acc-a\endcsname{77.1}
\expandafter\def\csname ppt@rev.desc.d2.sspro.p2.C.F.dual-d2.acc-a\endcsname{$+$77.1}
\expandafter\def\csname val@rev.desc.d2.sspro.p2.C.F.dual-d2.acc-b\endcsname{0.792}
\expandafter\def\csname cnt@rev.desc.d2.sspro.p2.C.F.dual-d2.acc-b\endcsname{1}
\expandafter\def\csname pct@rev.desc.d2.sspro.p2.C.F.dual-d2.acc-b\endcsname{79.2}
\expandafter\def\csname ppt@rev.desc.d2.sspro.p2.C.F.dual-d2.acc-b\endcsname{$+$79.2}
\expandafter\def\csname val@rev.desc.d2.sspro.p2.C.F.dual-d2.cond01\endcsname{0.545}
\expandafter\def\csname cnt@rev.desc.d2.sspro.p2.C.F.dual-d2.cond01\endcsname{1}
\expandafter\def\csname pct@rev.desc.d2.sspro.p2.C.F.dual-d2.cond01\endcsname{54.5}
\expandafter\def\csname ppt@rev.desc.d2.sspro.p2.C.F.dual-d2.cond01\endcsname{$+$54.5}
\expandafter\def\csname val@rev.desc.d2.sspro.p2.C.F.dual-d2.cond10\endcsname{0.135}
\expandafter\def\csname cnt@rev.desc.d2.sspro.p2.C.F.dual-d2.cond10\endcsname{0}
\expandafter\def\csname pct@rev.desc.d2.sspro.p2.C.F.dual-d2.cond10\endcsname{13.5}
\expandafter\def\csname ppt@rev.desc.d2.sspro.p2.C.F.dual-d2.cond10\endcsname{$+$13.5}
\expandafter\def\csname val@rev.desc.d2.sspro.p2.C.F.dual-d2.n\endcsname{48}
\expandafter\def\csname cnt@rev.desc.d2.sspro.p2.C.F.dual-d2.n\endcsname{48}
\expandafter\def\csname pct@rev.desc.d2.sspro.p2.C.F.dual-d2.n\endcsname{4800.0}
\expandafter\def\csname ppt@rev.desc.d2.sspro.p2.C.F.dual-d2.n\endcsname{$+$4800.0}
\expandafter\def\csname val@rev.desc.d2.sspro.p2.C.F.dual-d2.n-apps\endcsname{6}
\expandafter\def\csname cnt@rev.desc.d2.sspro.p2.C.F.dual-d2.n-apps\endcsname{6}
\expandafter\def\csname pct@rev.desc.d2.sspro.p2.C.F.dual-d2.n-apps\endcsname{600.0}
\expandafter\def\csname ppt@rev.desc.d2.sspro.p2.C.F.dual-d2.n-apps\endcsname{$+$600.0}
\expandafter\def\csname val@rev.desc.d2.sspro.p2.C.F.dual-d2.n01\endcsname{6}
\expandafter\def\csname cnt@rev.desc.d2.sspro.p2.C.F.dual-d2.n01\endcsname{6}
\expandafter\def\csname pct@rev.desc.d2.sspro.p2.C.F.dual-d2.n01\endcsname{600.0}
\expandafter\def\csname ppt@rev.desc.d2.sspro.p2.C.F.dual-d2.n01\endcsname{$+$600.0}
\expandafter\def\csname val@rev.desc.d2.sspro.p2.C.F.dual-d2.n10\endcsname{5}
\expandafter\def\csname cnt@rev.desc.d2.sspro.p2.C.F.dual-d2.n10\endcsname{5}
\expandafter\def\csname pct@rev.desc.d2.sspro.p2.C.F.dual-d2.n10\endcsname{500.0}
\expandafter\def\csname ppt@rev.desc.d2.sspro.p2.C.F.dual-d2.n10\endcsname{$+$500.0}
\expandafter\def\csname val@rev.desc.d2.sspro.p2.C.F.dual-d2.rate-any\endcsname{0.229}
\expandafter\def\csname cnt@rev.desc.d2.sspro.p2.C.F.dual-d2.rate-any\endcsname{0}
\expandafter\def\csname pct@rev.desc.d2.sspro.p2.C.F.dual-d2.rate-any\endcsname{22.9}
\expandafter\def\csname ppt@rev.desc.d2.sspro.p2.C.F.dual-d2.rate-any\endcsname{$+$22.9}
\expandafter\def\csname val@rev.desc.d2.sspro.p2.C.F.dual-d2.rate01\endcsname{0.125}
\expandafter\def\csname cnt@rev.desc.d2.sspro.p2.C.F.dual-d2.rate01\endcsname{0}
\expandafter\def\csname pct@rev.desc.d2.sspro.p2.C.F.dual-d2.rate01\endcsname{12.5}
\expandafter\def\csname ppt@rev.desc.d2.sspro.p2.C.F.dual-d2.rate01\endcsname{$+$12.5}
\expandafter\def\csname val@rev.desc.d2.sspro.p2.C.F.dual-d2.rate10\endcsname{0.104}
\expandafter\def\csname cnt@rev.desc.d2.sspro.p2.C.F.dual-d2.rate10\endcsname{0}
\expandafter\def\csname pct@rev.desc.d2.sspro.p2.C.F.dual-d2.rate10\endcsname{10.4}
\expandafter\def\csname ppt@rev.desc.d2.sspro.p2.C.F.dual-d2.rate10\endcsname{$+$10.4}
\expandafter\def\csname val@rev.desc.d2.sspro.p2.C.F.grid2-d3.acc-a\endcsname{0.792}
\expandafter\def\csname cnt@rev.desc.d2.sspro.p2.C.F.grid2-d3.acc-a\endcsname{1}
\expandafter\def\csname pct@rev.desc.d2.sspro.p2.C.F.grid2-d3.acc-a\endcsname{79.2}
\expandafter\def\csname ppt@rev.desc.d2.sspro.p2.C.F.grid2-d3.acc-a\endcsname{$+$79.2}
\expandafter\def\csname val@rev.desc.d2.sspro.p2.C.F.grid2-d3.acc-b\endcsname{0.833}
\expandafter\def\csname cnt@rev.desc.d2.sspro.p2.C.F.grid2-d3.acc-b\endcsname{1}
\expandafter\def\csname pct@rev.desc.d2.sspro.p2.C.F.grid2-d3.acc-b\endcsname{83.3}
\expandafter\def\csname ppt@rev.desc.d2.sspro.p2.C.F.grid2-d3.acc-b\endcsname{$+$83.3}
\expandafter\def\csname val@rev.desc.d2.sspro.p2.C.F.grid2-d3.cond01\endcsname{0.300}
\expandafter\def\csname cnt@rev.desc.d2.sspro.p2.C.F.grid2-d3.cond01\endcsname{0}
\expandafter\def\csname pct@rev.desc.d2.sspro.p2.C.F.grid2-d3.cond01\endcsname{30.0}
\expandafter\def\csname ppt@rev.desc.d2.sspro.p2.C.F.grid2-d3.cond01\endcsname{$+$30.0}
\expandafter\def\csname val@rev.desc.d2.sspro.p2.C.F.grid2-d3.cond10\endcsname{0.026}
\expandafter\def\csname cnt@rev.desc.d2.sspro.p2.C.F.grid2-d3.cond10\endcsname{0}
\expandafter\def\csname pct@rev.desc.d2.sspro.p2.C.F.grid2-d3.cond10\endcsname{2.6}
\expandafter\def\csname ppt@rev.desc.d2.sspro.p2.C.F.grid2-d3.cond10\endcsname{$+$2.6}
\expandafter\def\csname val@rev.desc.d2.sspro.p2.C.F.grid2-d3.n\endcsname{48}
\expandafter\def\csname cnt@rev.desc.d2.sspro.p2.C.F.grid2-d3.n\endcsname{48}
\expandafter\def\csname pct@rev.desc.d2.sspro.p2.C.F.grid2-d3.n\endcsname{4800.0}
\expandafter\def\csname ppt@rev.desc.d2.sspro.p2.C.F.grid2-d3.n\endcsname{$+$4800.0}
\expandafter\def\csname val@rev.desc.d2.sspro.p2.C.F.grid2-d3.n-apps\endcsname{6}
\expandafter\def\csname cnt@rev.desc.d2.sspro.p2.C.F.grid2-d3.n-apps\endcsname{6}
\expandafter\def\csname pct@rev.desc.d2.sspro.p2.C.F.grid2-d3.n-apps\endcsname{600.0}
\expandafter\def\csname ppt@rev.desc.d2.sspro.p2.C.F.grid2-d3.n-apps\endcsname{$+$600.0}
\expandafter\def\csname val@rev.desc.d2.sspro.p2.C.F.grid2-d3.n01\endcsname{3}
\expandafter\def\csname cnt@rev.desc.d2.sspro.p2.C.F.grid2-d3.n01\endcsname{3}
\expandafter\def\csname pct@rev.desc.d2.sspro.p2.C.F.grid2-d3.n01\endcsname{300.0}
\expandafter\def\csname ppt@rev.desc.d2.sspro.p2.C.F.grid2-d3.n01\endcsname{$+$300.0}
\expandafter\def\csname val@rev.desc.d2.sspro.p2.C.F.grid2-d3.n10\endcsname{1}
\expandafter\def\csname cnt@rev.desc.d2.sspro.p2.C.F.grid2-d3.n10\endcsname{1}
\expandafter\def\csname pct@rev.desc.d2.sspro.p2.C.F.grid2-d3.n10\endcsname{100.0}
\expandafter\def\csname ppt@rev.desc.d2.sspro.p2.C.F.grid2-d3.n10\endcsname{$+$100.0}
\expandafter\def\csname val@rev.desc.d2.sspro.p2.C.F.grid2-d3.rate-any\endcsname{0.083}
\expandafter\def\csname cnt@rev.desc.d2.sspro.p2.C.F.grid2-d3.rate-any\endcsname{0}
\expandafter\def\csname pct@rev.desc.d2.sspro.p2.C.F.grid2-d3.rate-any\endcsname{8.3}
\expandafter\def\csname ppt@rev.desc.d2.sspro.p2.C.F.grid2-d3.rate-any\endcsname{$+$8.3}
\expandafter\def\csname val@rev.desc.d2.sspro.p2.C.F.grid2-d3.rate01\endcsname{0.062}
\expandafter\def\csname cnt@rev.desc.d2.sspro.p2.C.F.grid2-d3.rate01\endcsname{0}
\expandafter\def\csname pct@rev.desc.d2.sspro.p2.C.F.grid2-d3.rate01\endcsname{6.2}
\expandafter\def\csname ppt@rev.desc.d2.sspro.p2.C.F.grid2-d3.rate01\endcsname{$+$6.2}
\expandafter\def\csname val@rev.desc.d2.sspro.p2.C.F.grid2-d3.rate10\endcsname{0.021}
\expandafter\def\csname cnt@rev.desc.d2.sspro.p2.C.F.grid2-d3.rate10\endcsname{0}
\expandafter\def\csname pct@rev.desc.d2.sspro.p2.C.F.grid2-d3.rate10\endcsname{2.1}
\expandafter\def\csname ppt@rev.desc.d2.sspro.p2.C.F.grid2-d3.rate10\endcsname{$+$2.1}
\expandafter\def\csname val@rev.desc.d2.sspro.p2.C.F.zoomclick-d3.acc-a\endcsname{0.854}
\expandafter\def\csname cnt@rev.desc.d2.sspro.p2.C.F.zoomclick-d3.acc-a\endcsname{1}
\expandafter\def\csname pct@rev.desc.d2.sspro.p2.C.F.zoomclick-d3.acc-a\endcsname{85.4}
\expandafter\def\csname ppt@rev.desc.d2.sspro.p2.C.F.zoomclick-d3.acc-a\endcsname{$+$85.4}
\expandafter\def\csname val@rev.desc.d2.sspro.p2.C.F.zoomclick-d3.acc-b\endcsname{0.896}
\expandafter\def\csname cnt@rev.desc.d2.sspro.p2.C.F.zoomclick-d3.acc-b\endcsname{1}
\expandafter\def\csname pct@rev.desc.d2.sspro.p2.C.F.zoomclick-d3.acc-b\endcsname{89.6}
\expandafter\def\csname ppt@rev.desc.d2.sspro.p2.C.F.zoomclick-d3.acc-b\endcsname{$+$89.6}
\expandafter\def\csname val@rev.desc.d2.sspro.p2.C.F.zoomclick-d3.cond01\endcsname{0.429}
\expandafter\def\csname cnt@rev.desc.d2.sspro.p2.C.F.zoomclick-d3.cond01\endcsname{0}
\expandafter\def\csname pct@rev.desc.d2.sspro.p2.C.F.zoomclick-d3.cond01\endcsname{42.9}
\expandafter\def\csname ppt@rev.desc.d2.sspro.p2.C.F.zoomclick-d3.cond01\endcsname{$+$42.9}
\expandafter\def\csname val@rev.desc.d2.sspro.p2.C.F.zoomclick-d3.cond10\endcsname{0.024}
\expandafter\def\csname cnt@rev.desc.d2.sspro.p2.C.F.zoomclick-d3.cond10\endcsname{0}
\expandafter\def\csname pct@rev.desc.d2.sspro.p2.C.F.zoomclick-d3.cond10\endcsname{2.4}
\expandafter\def\csname ppt@rev.desc.d2.sspro.p2.C.F.zoomclick-d3.cond10\endcsname{$+$2.4}
\expandafter\def\csname val@rev.desc.d2.sspro.p2.C.F.zoomclick-d3.n\endcsname{48}
\expandafter\def\csname cnt@rev.desc.d2.sspro.p2.C.F.zoomclick-d3.n\endcsname{48}
\expandafter\def\csname pct@rev.desc.d2.sspro.p2.C.F.zoomclick-d3.n\endcsname{4800.0}
\expandafter\def\csname ppt@rev.desc.d2.sspro.p2.C.F.zoomclick-d3.n\endcsname{$+$4800.0}
\expandafter\def\csname val@rev.desc.d2.sspro.p2.C.F.zoomclick-d3.n-apps\endcsname{6}
\expandafter\def\csname cnt@rev.desc.d2.sspro.p2.C.F.zoomclick-d3.n-apps\endcsname{6}
\expandafter\def\csname pct@rev.desc.d2.sspro.p2.C.F.zoomclick-d3.n-apps\endcsname{600.0}
\expandafter\def\csname ppt@rev.desc.d2.sspro.p2.C.F.zoomclick-d3.n-apps\endcsname{$+$600.0}
\expandafter\def\csname val@rev.desc.d2.sspro.p2.C.F.zoomclick-d3.n01\endcsname{3}
\expandafter\def\csname cnt@rev.desc.d2.sspro.p2.C.F.zoomclick-d3.n01\endcsname{3}
\expandafter\def\csname pct@rev.desc.d2.sspro.p2.C.F.zoomclick-d3.n01\endcsname{300.0}
\expandafter\def\csname ppt@rev.desc.d2.sspro.p2.C.F.zoomclick-d3.n01\endcsname{$+$300.0}
\expandafter\def\csname val@rev.desc.d2.sspro.p2.C.F.zoomclick-d3.n10\endcsname{1}
\expandafter\def\csname cnt@rev.desc.d2.sspro.p2.C.F.zoomclick-d3.n10\endcsname{1}
\expandafter\def\csname pct@rev.desc.d2.sspro.p2.C.F.zoomclick-d3.n10\endcsname{100.0}
\expandafter\def\csname ppt@rev.desc.d2.sspro.p2.C.F.zoomclick-d3.n10\endcsname{$+$100.0}
\expandafter\def\csname val@rev.desc.d2.sspro.p2.C.F.zoomclick-d3.rate-any\endcsname{0.083}
\expandafter\def\csname cnt@rev.desc.d2.sspro.p2.C.F.zoomclick-d3.rate-any\endcsname{0}
\expandafter\def\csname pct@rev.desc.d2.sspro.p2.C.F.zoomclick-d3.rate-any\endcsname{8.3}
\expandafter\def\csname ppt@rev.desc.d2.sspro.p2.C.F.zoomclick-d3.rate-any\endcsname{$+$8.3}
\expandafter\def\csname val@rev.desc.d2.sspro.p2.C.F.zoomclick-d3.rate01\endcsname{0.062}
\expandafter\def\csname cnt@rev.desc.d2.sspro.p2.C.F.zoomclick-d3.rate01\endcsname{0}
\expandafter\def\csname pct@rev.desc.d2.sspro.p2.C.F.zoomclick-d3.rate01\endcsname{6.2}
\expandafter\def\csname ppt@rev.desc.d2.sspro.p2.C.F.zoomclick-d3.rate01\endcsname{$+$6.2}
\expandafter\def\csname val@rev.desc.d2.sspro.p2.C.F.zoomclick-d3.rate10\endcsname{0.021}
\expandafter\def\csname cnt@rev.desc.d2.sspro.p2.C.F.zoomclick-d3.rate10\endcsname{0}
\expandafter\def\csname pct@rev.desc.d2.sspro.p2.C.F.zoomclick-d3.rate10\endcsname{2.1}
\expandafter\def\csname ppt@rev.desc.d2.sspro.p2.C.F.zoomclick-d3.rate10\endcsname{$+$2.1}
\expandafter\def\csname val@rev.desc.d2.sspro.p2.T.D.acc-a\endcsname{0.740}
\expandafter\def\csname cnt@rev.desc.d2.sspro.p2.T.D.acc-a\endcsname{1}
\expandafter\def\csname pct@rev.desc.d2.sspro.p2.T.D.acc-a\endcsname{74.0}
\expandafter\def\csname ppt@rev.desc.d2.sspro.p2.T.D.acc-a\endcsname{$+$74.0}
\expandafter\def\csname val@rev.desc.d2.sspro.p2.T.D.acc-b\endcsname{0.760}
\expandafter\def\csname cnt@rev.desc.d2.sspro.p2.T.D.acc-b\endcsname{1}
\expandafter\def\csname pct@rev.desc.d2.sspro.p2.T.D.acc-b\endcsname{76.0}
\expandafter\def\csname ppt@rev.desc.d2.sspro.p2.T.D.acc-b\endcsname{$+$76.0}
\expandafter\def\csname val@rev.desc.d2.sspro.p2.T.D.cond01\endcsname{0.320}
\expandafter\def\csname cnt@rev.desc.d2.sspro.p2.T.D.cond01\endcsname{0}
\expandafter\def\csname pct@rev.desc.d2.sspro.p2.T.D.cond01\endcsname{32.0}
\expandafter\def\csname ppt@rev.desc.d2.sspro.p2.T.D.cond01\endcsname{$+$32.0}
\expandafter\def\csname val@rev.desc.d2.sspro.p2.T.D.cond10\endcsname{0.085}
\expandafter\def\csname cnt@rev.desc.d2.sspro.p2.T.D.cond10\endcsname{0}
\expandafter\def\csname pct@rev.desc.d2.sspro.p2.T.D.cond10\endcsname{8.5}
\expandafter\def\csname ppt@rev.desc.d2.sspro.p2.T.D.cond10\endcsname{$+$8.5}
\expandafter\def\csname val@rev.desc.d2.sspro.p2.T.D.n\endcsname{96}
\expandafter\def\csname cnt@rev.desc.d2.sspro.p2.T.D.n\endcsname{96}
\expandafter\def\csname pct@rev.desc.d2.sspro.p2.T.D.n\endcsname{9600.0}
\expandafter\def\csname ppt@rev.desc.d2.sspro.p2.T.D.n\endcsname{$+$9600.0}
\expandafter\def\csname val@rev.desc.d2.sspro.p2.T.D.n-apps\endcsname{12}
\expandafter\def\csname cnt@rev.desc.d2.sspro.p2.T.D.n-apps\endcsname{12}
\expandafter\def\csname pct@rev.desc.d2.sspro.p2.T.D.n-apps\endcsname{1200.0}
\expandafter\def\csname ppt@rev.desc.d2.sspro.p2.T.D.n-apps\endcsname{$+$1200.0}
\expandafter\def\csname val@rev.desc.d2.sspro.p2.T.D.n01\endcsname{8}
\expandafter\def\csname cnt@rev.desc.d2.sspro.p2.T.D.n01\endcsname{8}
\expandafter\def\csname pct@rev.desc.d2.sspro.p2.T.D.n01\endcsname{800.0}
\expandafter\def\csname ppt@rev.desc.d2.sspro.p2.T.D.n01\endcsname{$+$800.0}
\expandafter\def\csname val@rev.desc.d2.sspro.p2.T.D.n10\endcsname{6}
\expandafter\def\csname cnt@rev.desc.d2.sspro.p2.T.D.n10\endcsname{6}
\expandafter\def\csname pct@rev.desc.d2.sspro.p2.T.D.n10\endcsname{600.0}
\expandafter\def\csname ppt@rev.desc.d2.sspro.p2.T.D.n10\endcsname{$+$600.0}
\expandafter\def\csname val@rev.desc.d2.sspro.p2.T.D.rate-any\endcsname{0.146}
\expandafter\def\csname cnt@rev.desc.d2.sspro.p2.T.D.rate-any\endcsname{0}
\expandafter\def\csname pct@rev.desc.d2.sspro.p2.T.D.rate-any\endcsname{14.6}
\expandafter\def\csname ppt@rev.desc.d2.sspro.p2.T.D.rate-any\endcsname{$+$14.6}
\expandafter\def\csname val@rev.desc.d2.sspro.p2.T.D.rate01\endcsname{0.083}
\expandafter\def\csname cnt@rev.desc.d2.sspro.p2.T.D.rate01\endcsname{0}
\expandafter\def\csname pct@rev.desc.d2.sspro.p2.T.D.rate01\endcsname{8.3}
\expandafter\def\csname ppt@rev.desc.d2.sspro.p2.T.D.rate01\endcsname{$+$8.3}
\expandafter\def\csname val@rev.desc.d2.sspro.p2.T.D.rate10\endcsname{0.062}
\expandafter\def\csname cnt@rev.desc.d2.sspro.p2.T.D.rate10\endcsname{0}
\expandafter\def\csname pct@rev.desc.d2.sspro.p2.T.D.rate10\endcsname{6.2}
\expandafter\def\csname ppt@rev.desc.d2.sspro.p2.T.D.rate10\endcsname{$+$6.2}
\expandafter\def\csname val@rev.desc.d2.sspro.p2.T.F.center-d1.acc-a\endcsname{0.844}
\expandafter\def\csname cnt@rev.desc.d2.sspro.p2.T.F.center-d1.acc-a\endcsname{1}
\expandafter\def\csname pct@rev.desc.d2.sspro.p2.T.F.center-d1.acc-a\endcsname{84.4}
\expandafter\def\csname ppt@rev.desc.d2.sspro.p2.T.F.center-d1.acc-a\endcsname{$+$84.4}
\expandafter\def\csname val@rev.desc.d2.sspro.p2.T.F.center-d1.acc-b\endcsname{0.917}
\expandafter\def\csname cnt@rev.desc.d2.sspro.p2.T.F.center-d1.acc-b\endcsname{1}
\expandafter\def\csname pct@rev.desc.d2.sspro.p2.T.F.center-d1.acc-b\endcsname{91.7}
\expandafter\def\csname ppt@rev.desc.d2.sspro.p2.T.F.center-d1.acc-b\endcsname{$+$91.7}
\expandafter\def\csname val@rev.desc.d2.sspro.p2.T.F.center-d1.cond01\endcsname{0.533}
\expandafter\def\csname cnt@rev.desc.d2.sspro.p2.T.F.center-d1.cond01\endcsname{1}
\expandafter\def\csname pct@rev.desc.d2.sspro.p2.T.F.center-d1.cond01\endcsname{53.3}
\expandafter\def\csname ppt@rev.desc.d2.sspro.p2.T.F.center-d1.cond01\endcsname{$+$53.3}
\expandafter\def\csname val@rev.desc.d2.sspro.p2.T.F.center-d1.cond10\endcsname{0.012}
\expandafter\def\csname cnt@rev.desc.d2.sspro.p2.T.F.center-d1.cond10\endcsname{0}
\expandafter\def\csname pct@rev.desc.d2.sspro.p2.T.F.center-d1.cond10\endcsname{1.2}
\expandafter\def\csname ppt@rev.desc.d2.sspro.p2.T.F.center-d1.cond10\endcsname{$+$1.2}
\expandafter\def\csname val@rev.desc.d2.sspro.p2.T.F.center-d1.n\endcsname{96}
\expandafter\def\csname cnt@rev.desc.d2.sspro.p2.T.F.center-d1.n\endcsname{96}
\expandafter\def\csname pct@rev.desc.d2.sspro.p2.T.F.center-d1.n\endcsname{9600.0}
\expandafter\def\csname ppt@rev.desc.d2.sspro.p2.T.F.center-d1.n\endcsname{$+$9600.0}
\expandafter\def\csname val@rev.desc.d2.sspro.p2.T.F.center-d1.n-apps\endcsname{12}
\expandafter\def\csname cnt@rev.desc.d2.sspro.p2.T.F.center-d1.n-apps\endcsname{12}
\expandafter\def\csname pct@rev.desc.d2.sspro.p2.T.F.center-d1.n-apps\endcsname{1200.0}
\expandafter\def\csname ppt@rev.desc.d2.sspro.p2.T.F.center-d1.n-apps\endcsname{$+$1200.0}
\expandafter\def\csname val@rev.desc.d2.sspro.p2.T.F.center-d1.n01\endcsname{8}
\expandafter\def\csname cnt@rev.desc.d2.sspro.p2.T.F.center-d1.n01\endcsname{8}
\expandafter\def\csname pct@rev.desc.d2.sspro.p2.T.F.center-d1.n01\endcsname{800.0}
\expandafter\def\csname ppt@rev.desc.d2.sspro.p2.T.F.center-d1.n01\endcsname{$+$800.0}
\expandafter\def\csname val@rev.desc.d2.sspro.p2.T.F.center-d1.n10\endcsname{1}
\expandafter\def\csname cnt@rev.desc.d2.sspro.p2.T.F.center-d1.n10\endcsname{1}
\expandafter\def\csname pct@rev.desc.d2.sspro.p2.T.F.center-d1.n10\endcsname{100.0}
\expandafter\def\csname ppt@rev.desc.d2.sspro.p2.T.F.center-d1.n10\endcsname{$+$100.0}
\expandafter\def\csname val@rev.desc.d2.sspro.p2.T.F.center-d1.rate-any\endcsname{0.094}
\expandafter\def\csname cnt@rev.desc.d2.sspro.p2.T.F.center-d1.rate-any\endcsname{0}
\expandafter\def\csname pct@rev.desc.d2.sspro.p2.T.F.center-d1.rate-any\endcsname{9.4}
\expandafter\def\csname ppt@rev.desc.d2.sspro.p2.T.F.center-d1.rate-any\endcsname{$+$9.4}
\expandafter\def\csname val@rev.desc.d2.sspro.p2.T.F.center-d1.rate01\endcsname{0.083}
\expandafter\def\csname cnt@rev.desc.d2.sspro.p2.T.F.center-d1.rate01\endcsname{0}
\expandafter\def\csname pct@rev.desc.d2.sspro.p2.T.F.center-d1.rate01\endcsname{8.3}
\expandafter\def\csname ppt@rev.desc.d2.sspro.p2.T.F.center-d1.rate01\endcsname{$+$8.3}
\expandafter\def\csname val@rev.desc.d2.sspro.p2.T.F.center-d1.rate10\endcsname{0.010}
\expandafter\def\csname cnt@rev.desc.d2.sspro.p2.T.F.center-d1.rate10\endcsname{0}
\expandafter\def\csname pct@rev.desc.d2.sspro.p2.T.F.center-d1.rate10\endcsname{1.0}
\expandafter\def\csname ppt@rev.desc.d2.sspro.p2.T.F.center-d1.rate10\endcsname{$+$1.0}
\expandafter\def\csname val@rev.desc.d2.sspro.p2.T.F.context-d1.acc-a\endcsname{0.854}
\expandafter\def\csname cnt@rev.desc.d2.sspro.p2.T.F.context-d1.acc-a\endcsname{1}
\expandafter\def\csname pct@rev.desc.d2.sspro.p2.T.F.context-d1.acc-a\endcsname{85.4}
\expandafter\def\csname ppt@rev.desc.d2.sspro.p2.T.F.context-d1.acc-a\endcsname{$+$85.4}
\expandafter\def\csname val@rev.desc.d2.sspro.p2.T.F.context-d1.acc-b\endcsname{0.885}
\expandafter\def\csname cnt@rev.desc.d2.sspro.p2.T.F.context-d1.acc-b\endcsname{1}
\expandafter\def\csname pct@rev.desc.d2.sspro.p2.T.F.context-d1.acc-b\endcsname{88.5}
\expandafter\def\csname ppt@rev.desc.d2.sspro.p2.T.F.context-d1.acc-b\endcsname{$+$88.5}
\expandafter\def\csname val@rev.desc.d2.sspro.p2.T.F.context-d1.cond01\endcsname{0.357}
\expandafter\def\csname cnt@rev.desc.d2.sspro.p2.T.F.context-d1.cond01\endcsname{0}
\expandafter\def\csname pct@rev.desc.d2.sspro.p2.T.F.context-d1.cond01\endcsname{35.7}
\expandafter\def\csname ppt@rev.desc.d2.sspro.p2.T.F.context-d1.cond01\endcsname{$+$35.7}
\expandafter\def\csname val@rev.desc.d2.sspro.p2.T.F.context-d1.cond10\endcsname{0.024}
\expandafter\def\csname cnt@rev.desc.d2.sspro.p2.T.F.context-d1.cond10\endcsname{0}
\expandafter\def\csname pct@rev.desc.d2.sspro.p2.T.F.context-d1.cond10\endcsname{2.4}
\expandafter\def\csname ppt@rev.desc.d2.sspro.p2.T.F.context-d1.cond10\endcsname{$+$2.4}
\expandafter\def\csname val@rev.desc.d2.sspro.p2.T.F.context-d1.n\endcsname{96}
\expandafter\def\csname cnt@rev.desc.d2.sspro.p2.T.F.context-d1.n\endcsname{96}
\expandafter\def\csname pct@rev.desc.d2.sspro.p2.T.F.context-d1.n\endcsname{9600.0}
\expandafter\def\csname ppt@rev.desc.d2.sspro.p2.T.F.context-d1.n\endcsname{$+$9600.0}
\expandafter\def\csname val@rev.desc.d2.sspro.p2.T.F.context-d1.n-apps\endcsname{12}
\expandafter\def\csname cnt@rev.desc.d2.sspro.p2.T.F.context-d1.n-apps\endcsname{12}
\expandafter\def\csname pct@rev.desc.d2.sspro.p2.T.F.context-d1.n-apps\endcsname{1200.0}
\expandafter\def\csname ppt@rev.desc.d2.sspro.p2.T.F.context-d1.n-apps\endcsname{$+$1200.0}
\expandafter\def\csname val@rev.desc.d2.sspro.p2.T.F.context-d1.n01\endcsname{5}
\expandafter\def\csname cnt@rev.desc.d2.sspro.p2.T.F.context-d1.n01\endcsname{5}
\expandafter\def\csname pct@rev.desc.d2.sspro.p2.T.F.context-d1.n01\endcsname{500.0}
\expandafter\def\csname ppt@rev.desc.d2.sspro.p2.T.F.context-d1.n01\endcsname{$+$500.0}
\expandafter\def\csname val@rev.desc.d2.sspro.p2.T.F.context-d1.n10\endcsname{2}
\expandafter\def\csname cnt@rev.desc.d2.sspro.p2.T.F.context-d1.n10\endcsname{2}
\expandafter\def\csname pct@rev.desc.d2.sspro.p2.T.F.context-d1.n10\endcsname{200.0}
\expandafter\def\csname ppt@rev.desc.d2.sspro.p2.T.F.context-d1.n10\endcsname{$+$200.0}
\expandafter\def\csname val@rev.desc.d2.sspro.p2.T.F.context-d1.rate-any\endcsname{0.073}
\expandafter\def\csname cnt@rev.desc.d2.sspro.p2.T.F.context-d1.rate-any\endcsname{0}
\expandafter\def\csname pct@rev.desc.d2.sspro.p2.T.F.context-d1.rate-any\endcsname{7.3}
\expandafter\def\csname ppt@rev.desc.d2.sspro.p2.T.F.context-d1.rate-any\endcsname{$+$7.3}
\expandafter\def\csname val@rev.desc.d2.sspro.p2.T.F.context-d1.rate01\endcsname{0.052}
\expandafter\def\csname cnt@rev.desc.d2.sspro.p2.T.F.context-d1.rate01\endcsname{0}
\expandafter\def\csname pct@rev.desc.d2.sspro.p2.T.F.context-d1.rate01\endcsname{5.2}
\expandafter\def\csname ppt@rev.desc.d2.sspro.p2.T.F.context-d1.rate01\endcsname{$+$5.2}
\expandafter\def\csname val@rev.desc.d2.sspro.p2.T.F.context-d1.rate10\endcsname{0.021}
\expandafter\def\csname cnt@rev.desc.d2.sspro.p2.T.F.context-d1.rate10\endcsname{0}
\expandafter\def\csname pct@rev.desc.d2.sspro.p2.T.F.context-d1.rate10\endcsname{2.1}
\expandafter\def\csname ppt@rev.desc.d2.sspro.p2.T.F.context-d1.rate10\endcsname{$+$2.1}
\expandafter\def\csname val@rev.desc.d2.sspro.p2.T.F.dual-d2.acc-a\endcsname{0.885}
\expandafter\def\csname cnt@rev.desc.d2.sspro.p2.T.F.dual-d2.acc-a\endcsname{1}
\expandafter\def\csname pct@rev.desc.d2.sspro.p2.T.F.dual-d2.acc-a\endcsname{88.5}
\expandafter\def\csname ppt@rev.desc.d2.sspro.p2.T.F.dual-d2.acc-a\endcsname{$+$88.5}
\expandafter\def\csname val@rev.desc.d2.sspro.p2.T.F.dual-d2.acc-b\endcsname{0.917}
\expandafter\def\csname cnt@rev.desc.d2.sspro.p2.T.F.dual-d2.acc-b\endcsname{1}
\expandafter\def\csname pct@rev.desc.d2.sspro.p2.T.F.dual-d2.acc-b\endcsname{91.7}
\expandafter\def\csname ppt@rev.desc.d2.sspro.p2.T.F.dual-d2.acc-b\endcsname{$+$91.7}
\expandafter\def\csname val@rev.desc.d2.sspro.p2.T.F.dual-d2.cond01\endcsname{0.545}
\expandafter\def\csname cnt@rev.desc.d2.sspro.p2.T.F.dual-d2.cond01\endcsname{1}
\expandafter\def\csname pct@rev.desc.d2.sspro.p2.T.F.dual-d2.cond01\endcsname{54.5}
\expandafter\def\csname ppt@rev.desc.d2.sspro.p2.T.F.dual-d2.cond01\endcsname{$+$54.5}
\expandafter\def\csname val@rev.desc.d2.sspro.p2.T.F.dual-d2.cond10\endcsname{0.035}
\expandafter\def\csname cnt@rev.desc.d2.sspro.p2.T.F.dual-d2.cond10\endcsname{0}
\expandafter\def\csname pct@rev.desc.d2.sspro.p2.T.F.dual-d2.cond10\endcsname{3.5}
\expandafter\def\csname ppt@rev.desc.d2.sspro.p2.T.F.dual-d2.cond10\endcsname{$+$3.5}
\expandafter\def\csname val@rev.desc.d2.sspro.p2.T.F.dual-d2.n\endcsname{96}
\expandafter\def\csname cnt@rev.desc.d2.sspro.p2.T.F.dual-d2.n\endcsname{96}
\expandafter\def\csname pct@rev.desc.d2.sspro.p2.T.F.dual-d2.n\endcsname{9600.0}
\expandafter\def\csname ppt@rev.desc.d2.sspro.p2.T.F.dual-d2.n\endcsname{$+$9600.0}
\expandafter\def\csname val@rev.desc.d2.sspro.p2.T.F.dual-d2.n-apps\endcsname{12}
\expandafter\def\csname cnt@rev.desc.d2.sspro.p2.T.F.dual-d2.n-apps\endcsname{12}
\expandafter\def\csname pct@rev.desc.d2.sspro.p2.T.F.dual-d2.n-apps\endcsname{1200.0}
\expandafter\def\csname ppt@rev.desc.d2.sspro.p2.T.F.dual-d2.n-apps\endcsname{$+$1200.0}
\expandafter\def\csname val@rev.desc.d2.sspro.p2.T.F.dual-d2.n01\endcsname{6}
\expandafter\def\csname cnt@rev.desc.d2.sspro.p2.T.F.dual-d2.n01\endcsname{6}
\expandafter\def\csname pct@rev.desc.d2.sspro.p2.T.F.dual-d2.n01\endcsname{600.0}
\expandafter\def\csname ppt@rev.desc.d2.sspro.p2.T.F.dual-d2.n01\endcsname{$+$600.0}
\expandafter\def\csname val@rev.desc.d2.sspro.p2.T.F.dual-d2.n10\endcsname{3}
\expandafter\def\csname cnt@rev.desc.d2.sspro.p2.T.F.dual-d2.n10\endcsname{3}
\expandafter\def\csname pct@rev.desc.d2.sspro.p2.T.F.dual-d2.n10\endcsname{300.0}
\expandafter\def\csname ppt@rev.desc.d2.sspro.p2.T.F.dual-d2.n10\endcsname{$+$300.0}
\expandafter\def\csname val@rev.desc.d2.sspro.p2.T.F.dual-d2.rate-any\endcsname{0.094}
\expandafter\def\csname cnt@rev.desc.d2.sspro.p2.T.F.dual-d2.rate-any\endcsname{0}
\expandafter\def\csname pct@rev.desc.d2.sspro.p2.T.F.dual-d2.rate-any\endcsname{9.4}
\expandafter\def\csname ppt@rev.desc.d2.sspro.p2.T.F.dual-d2.rate-any\endcsname{$+$9.4}
\expandafter\def\csname val@rev.desc.d2.sspro.p2.T.F.dual-d2.rate01\endcsname{0.062}
\expandafter\def\csname cnt@rev.desc.d2.sspro.p2.T.F.dual-d2.rate01\endcsname{0}
\expandafter\def\csname pct@rev.desc.d2.sspro.p2.T.F.dual-d2.rate01\endcsname{6.2}
\expandafter\def\csname ppt@rev.desc.d2.sspro.p2.T.F.dual-d2.rate01\endcsname{$+$6.2}
\expandafter\def\csname val@rev.desc.d2.sspro.p2.T.F.dual-d2.rate10\endcsname{0.031}
\expandafter\def\csname cnt@rev.desc.d2.sspro.p2.T.F.dual-d2.rate10\endcsname{0}
\expandafter\def\csname pct@rev.desc.d2.sspro.p2.T.F.dual-d2.rate10\endcsname{3.1}
\expandafter\def\csname ppt@rev.desc.d2.sspro.p2.T.F.dual-d2.rate10\endcsname{$+$3.1}
\expandafter\def\csname val@rev.desc.d2.sspro.p2.T.F.grid2-d3.acc-a\endcsname{0.854}
\expandafter\def\csname cnt@rev.desc.d2.sspro.p2.T.F.grid2-d3.acc-a\endcsname{1}
\expandafter\def\csname pct@rev.desc.d2.sspro.p2.T.F.grid2-d3.acc-a\endcsname{85.4}
\expandafter\def\csname ppt@rev.desc.d2.sspro.p2.T.F.grid2-d3.acc-a\endcsname{$+$85.4}
\expandafter\def\csname val@rev.desc.d2.sspro.p2.T.F.grid2-d3.acc-b\endcsname{0.906}
\expandafter\def\csname cnt@rev.desc.d2.sspro.p2.T.F.grid2-d3.acc-b\endcsname{1}
\expandafter\def\csname pct@rev.desc.d2.sspro.p2.T.F.grid2-d3.acc-b\endcsname{90.6}
\expandafter\def\csname ppt@rev.desc.d2.sspro.p2.T.F.grid2-d3.acc-b\endcsname{$+$90.6}
\expandafter\def\csname val@rev.desc.d2.sspro.p2.T.F.grid2-d3.cond01\endcsname{0.643}
\expandafter\def\csname cnt@rev.desc.d2.sspro.p2.T.F.grid2-d3.cond01\endcsname{1}
\expandafter\def\csname pct@rev.desc.d2.sspro.p2.T.F.grid2-d3.cond01\endcsname{64.3}
\expandafter\def\csname ppt@rev.desc.d2.sspro.p2.T.F.grid2-d3.cond01\endcsname{$+$64.3}
\expandafter\def\csname val@rev.desc.d2.sspro.p2.T.F.grid2-d3.cond10\endcsname{0.049}
\expandafter\def\csname cnt@rev.desc.d2.sspro.p2.T.F.grid2-d3.cond10\endcsname{0}
\expandafter\def\csname pct@rev.desc.d2.sspro.p2.T.F.grid2-d3.cond10\endcsname{4.9}
\expandafter\def\csname ppt@rev.desc.d2.sspro.p2.T.F.grid2-d3.cond10\endcsname{$+$4.9}
\expandafter\def\csname val@rev.desc.d2.sspro.p2.T.F.grid2-d3.n\endcsname{96}
\expandafter\def\csname cnt@rev.desc.d2.sspro.p2.T.F.grid2-d3.n\endcsname{96}
\expandafter\def\csname pct@rev.desc.d2.sspro.p2.T.F.grid2-d3.n\endcsname{9600.0}
\expandafter\def\csname ppt@rev.desc.d2.sspro.p2.T.F.grid2-d3.n\endcsname{$+$9600.0}
\expandafter\def\csname val@rev.desc.d2.sspro.p2.T.F.grid2-d3.n-apps\endcsname{12}
\expandafter\def\csname cnt@rev.desc.d2.sspro.p2.T.F.grid2-d3.n-apps\endcsname{12}
\expandafter\def\csname pct@rev.desc.d2.sspro.p2.T.F.grid2-d3.n-apps\endcsname{1200.0}
\expandafter\def\csname ppt@rev.desc.d2.sspro.p2.T.F.grid2-d3.n-apps\endcsname{$+$1200.0}
\expandafter\def\csname val@rev.desc.d2.sspro.p2.T.F.grid2-d3.n01\endcsname{9}
\expandafter\def\csname cnt@rev.desc.d2.sspro.p2.T.F.grid2-d3.n01\endcsname{9}
\expandafter\def\csname pct@rev.desc.d2.sspro.p2.T.F.grid2-d3.n01\endcsname{900.0}
\expandafter\def\csname ppt@rev.desc.d2.sspro.p2.T.F.grid2-d3.n01\endcsname{$+$900.0}
\expandafter\def\csname val@rev.desc.d2.sspro.p2.T.F.grid2-d3.n10\endcsname{4}
\expandafter\def\csname cnt@rev.desc.d2.sspro.p2.T.F.grid2-d3.n10\endcsname{4}
\expandafter\def\csname pct@rev.desc.d2.sspro.p2.T.F.grid2-d3.n10\endcsname{400.0}
\expandafter\def\csname ppt@rev.desc.d2.sspro.p2.T.F.grid2-d3.n10\endcsname{$+$400.0}
\expandafter\def\csname val@rev.desc.d2.sspro.p2.T.F.grid2-d3.rate-any\endcsname{0.135}
\expandafter\def\csname cnt@rev.desc.d2.sspro.p2.T.F.grid2-d3.rate-any\endcsname{0}
\expandafter\def\csname pct@rev.desc.d2.sspro.p2.T.F.grid2-d3.rate-any\endcsname{13.5}
\expandafter\def\csname ppt@rev.desc.d2.sspro.p2.T.F.grid2-d3.rate-any\endcsname{$+$13.5}
\expandafter\def\csname val@rev.desc.d2.sspro.p2.T.F.grid2-d3.rate01\endcsname{0.094}
\expandafter\def\csname cnt@rev.desc.d2.sspro.p2.T.F.grid2-d3.rate01\endcsname{0}
\expandafter\def\csname pct@rev.desc.d2.sspro.p2.T.F.grid2-d3.rate01\endcsname{9.4}
\expandafter\def\csname ppt@rev.desc.d2.sspro.p2.T.F.grid2-d3.rate01\endcsname{$+$9.4}
\expandafter\def\csname val@rev.desc.d2.sspro.p2.T.F.grid2-d3.rate10\endcsname{0.042}
\expandafter\def\csname cnt@rev.desc.d2.sspro.p2.T.F.grid2-d3.rate10\endcsname{0}
\expandafter\def\csname pct@rev.desc.d2.sspro.p2.T.F.grid2-d3.rate10\endcsname{4.2}
\expandafter\def\csname ppt@rev.desc.d2.sspro.p2.T.F.grid2-d3.rate10\endcsname{$+$4.2}
\expandafter\def\csname val@rev.desc.d2.sspro.p2.T.F.zoomclick-d3.acc-a\endcsname{0.917}
\expandafter\def\csname cnt@rev.desc.d2.sspro.p2.T.F.zoomclick-d3.acc-a\endcsname{1}
\expandafter\def\csname pct@rev.desc.d2.sspro.p2.T.F.zoomclick-d3.acc-a\endcsname{91.7}
\expandafter\def\csname ppt@rev.desc.d2.sspro.p2.T.F.zoomclick-d3.acc-a\endcsname{$+$91.7}
\expandafter\def\csname val@rev.desc.d2.sspro.p2.T.F.zoomclick-d3.acc-b\endcsname{0.927}
\expandafter\def\csname cnt@rev.desc.d2.sspro.p2.T.F.zoomclick-d3.acc-b\endcsname{1}
\expandafter\def\csname pct@rev.desc.d2.sspro.p2.T.F.zoomclick-d3.acc-b\endcsname{92.7}
\expandafter\def\csname ppt@rev.desc.d2.sspro.p2.T.F.zoomclick-d3.acc-b\endcsname{$+$92.7}
\expandafter\def\csname val@rev.desc.d2.sspro.p2.T.F.zoomclick-d3.cond01\endcsname{0.375}
\expandafter\def\csname cnt@rev.desc.d2.sspro.p2.T.F.zoomclick-d3.cond01\endcsname{0}
\expandafter\def\csname pct@rev.desc.d2.sspro.p2.T.F.zoomclick-d3.cond01\endcsname{37.5}
\expandafter\def\csname ppt@rev.desc.d2.sspro.p2.T.F.zoomclick-d3.cond01\endcsname{$+$37.5}
\expandafter\def\csname val@rev.desc.d2.sspro.p2.T.F.zoomclick-d3.cond10\endcsname{0.023}
\expandafter\def\csname cnt@rev.desc.d2.sspro.p2.T.F.zoomclick-d3.cond10\endcsname{0}
\expandafter\def\csname pct@rev.desc.d2.sspro.p2.T.F.zoomclick-d3.cond10\endcsname{2.3}
\expandafter\def\csname ppt@rev.desc.d2.sspro.p2.T.F.zoomclick-d3.cond10\endcsname{$+$2.3}
\expandafter\def\csname val@rev.desc.d2.sspro.p2.T.F.zoomclick-d3.n\endcsname{96}
\expandafter\def\csname cnt@rev.desc.d2.sspro.p2.T.F.zoomclick-d3.n\endcsname{96}
\expandafter\def\csname pct@rev.desc.d2.sspro.p2.T.F.zoomclick-d3.n\endcsname{9600.0}
\expandafter\def\csname ppt@rev.desc.d2.sspro.p2.T.F.zoomclick-d3.n\endcsname{$+$9600.0}
\expandafter\def\csname val@rev.desc.d2.sspro.p2.T.F.zoomclick-d3.n-apps\endcsname{12}
\expandafter\def\csname cnt@rev.desc.d2.sspro.p2.T.F.zoomclick-d3.n-apps\endcsname{12}
\expandafter\def\csname pct@rev.desc.d2.sspro.p2.T.F.zoomclick-d3.n-apps\endcsname{1200.0}
\expandafter\def\csname ppt@rev.desc.d2.sspro.p2.T.F.zoomclick-d3.n-apps\endcsname{$+$1200.0}
\expandafter\def\csname val@rev.desc.d2.sspro.p2.T.F.zoomclick-d3.n01\endcsname{3}
\expandafter\def\csname cnt@rev.desc.d2.sspro.p2.T.F.zoomclick-d3.n01\endcsname{3}
\expandafter\def\csname pct@rev.desc.d2.sspro.p2.T.F.zoomclick-d3.n01\endcsname{300.0}
\expandafter\def\csname ppt@rev.desc.d2.sspro.p2.T.F.zoomclick-d3.n01\endcsname{$+$300.0}
\expandafter\def\csname val@rev.desc.d2.sspro.p2.T.F.zoomclick-d3.n10\endcsname{2}
\expandafter\def\csname cnt@rev.desc.d2.sspro.p2.T.F.zoomclick-d3.n10\endcsname{2}
\expandafter\def\csname pct@rev.desc.d2.sspro.p2.T.F.zoomclick-d3.n10\endcsname{200.0}
\expandafter\def\csname ppt@rev.desc.d2.sspro.p2.T.F.zoomclick-d3.n10\endcsname{$+$200.0}
\expandafter\def\csname val@rev.desc.d2.sspro.p2.T.F.zoomclick-d3.rate-any\endcsname{0.052}
\expandafter\def\csname cnt@rev.desc.d2.sspro.p2.T.F.zoomclick-d3.rate-any\endcsname{0}
\expandafter\def\csname pct@rev.desc.d2.sspro.p2.T.F.zoomclick-d3.rate-any\endcsname{5.2}
\expandafter\def\csname ppt@rev.desc.d2.sspro.p2.T.F.zoomclick-d3.rate-any\endcsname{$+$5.2}
\expandafter\def\csname val@rev.desc.d2.sspro.p2.T.F.zoomclick-d3.rate01\endcsname{0.031}
\expandafter\def\csname cnt@rev.desc.d2.sspro.p2.T.F.zoomclick-d3.rate01\endcsname{0}
\expandafter\def\csname pct@rev.desc.d2.sspro.p2.T.F.zoomclick-d3.rate01\endcsname{3.1}
\expandafter\def\csname ppt@rev.desc.d2.sspro.p2.T.F.zoomclick-d3.rate01\endcsname{$+$3.1}
\expandafter\def\csname val@rev.desc.d2.sspro.p2.T.F.zoomclick-d3.rate10\endcsname{0.021}
\expandafter\def\csname cnt@rev.desc.d2.sspro.p2.T.F.zoomclick-d3.rate10\endcsname{0}
\expandafter\def\csname pct@rev.desc.d2.sspro.p2.T.F.zoomclick-d3.rate10\endcsname{2.1}
\expandafter\def\csname ppt@rev.desc.d2.sspro.p2.T.F.zoomclick-d3.rate10\endcsname{$+$2.1}
\expandafter\def\csname val@rev.desc.d2.sspro.rep.haiku.C.D.acc-a\endcsname{0.667}
\expandafter\def\csname cnt@rev.desc.d2.sspro.rep.haiku.C.D.acc-a\endcsname{1}
\expandafter\def\csname pct@rev.desc.d2.sspro.rep.haiku.C.D.acc-a\endcsname{66.7}
\expandafter\def\csname ppt@rev.desc.d2.sspro.rep.haiku.C.D.acc-a\endcsname{$+$66.7}
\expandafter\def\csname val@rev.desc.d2.sspro.rep.haiku.C.D.acc-b\endcsname{0.600}
\expandafter\def\csname cnt@rev.desc.d2.sspro.rep.haiku.C.D.acc-b\endcsname{1}
\expandafter\def\csname pct@rev.desc.d2.sspro.rep.haiku.C.D.acc-b\endcsname{60.0}
\expandafter\def\csname ppt@rev.desc.d2.sspro.rep.haiku.C.D.acc-b\endcsname{$+$60.0}
\expandafter\def\csname val@rev.desc.d2.sspro.rep.haiku.C.D.cond01\endcsname{0.000}
\expandafter\def\csname cnt@rev.desc.d2.sspro.rep.haiku.C.D.cond01\endcsname{0}
\expandafter\def\csname pct@rev.desc.d2.sspro.rep.haiku.C.D.cond01\endcsname{0.0}
\expandafter\def\csname ppt@rev.desc.d2.sspro.rep.haiku.C.D.cond01\endcsname{0.0}
\expandafter\def\csname val@rev.desc.d2.sspro.rep.haiku.C.D.cond10\endcsname{0.100}
\expandafter\def\csname cnt@rev.desc.d2.sspro.rep.haiku.C.D.cond10\endcsname{0}
\expandafter\def\csname pct@rev.desc.d2.sspro.rep.haiku.C.D.cond10\endcsname{10.0}
\expandafter\def\csname ppt@rev.desc.d2.sspro.rep.haiku.C.D.cond10\endcsname{$+$10.0}
\expandafter\def\csname val@rev.desc.d2.sspro.rep.haiku.C.D.n\endcsname{15}
\expandafter\def\csname cnt@rev.desc.d2.sspro.rep.haiku.C.D.n\endcsname{15}
\expandafter\def\csname pct@rev.desc.d2.sspro.rep.haiku.C.D.n\endcsname{1500.0}
\expandafter\def\csname ppt@rev.desc.d2.sspro.rep.haiku.C.D.n\endcsname{$+$1500.0}
\expandafter\def\csname val@rev.desc.d2.sspro.rep.haiku.C.D.n-apps\endcsname{6}
\expandafter\def\csname cnt@rev.desc.d2.sspro.rep.haiku.C.D.n-apps\endcsname{6}
\expandafter\def\csname pct@rev.desc.d2.sspro.rep.haiku.C.D.n-apps\endcsname{600.0}
\expandafter\def\csname ppt@rev.desc.d2.sspro.rep.haiku.C.D.n-apps\endcsname{$+$600.0}
\expandafter\def\csname val@rev.desc.d2.sspro.rep.haiku.C.D.n01\endcsname{0}
\expandafter\def\csname cnt@rev.desc.d2.sspro.rep.haiku.C.D.n01\endcsname{0}
\expandafter\def\csname pct@rev.desc.d2.sspro.rep.haiku.C.D.n01\endcsname{0.0}
\expandafter\def\csname ppt@rev.desc.d2.sspro.rep.haiku.C.D.n01\endcsname{0.0}
\expandafter\def\csname val@rev.desc.d2.sspro.rep.haiku.C.D.n10\endcsname{1}
\expandafter\def\csname cnt@rev.desc.d2.sspro.rep.haiku.C.D.n10\endcsname{1}
\expandafter\def\csname pct@rev.desc.d2.sspro.rep.haiku.C.D.n10\endcsname{100.0}
\expandafter\def\csname ppt@rev.desc.d2.sspro.rep.haiku.C.D.n10\endcsname{$+$100.0}
\expandafter\def\csname val@rev.desc.d2.sspro.rep.haiku.C.D.rate-any\endcsname{0.067}
\expandafter\def\csname cnt@rev.desc.d2.sspro.rep.haiku.C.D.rate-any\endcsname{0}
\expandafter\def\csname pct@rev.desc.d2.sspro.rep.haiku.C.D.rate-any\endcsname{6.7}
\expandafter\def\csname ppt@rev.desc.d2.sspro.rep.haiku.C.D.rate-any\endcsname{$+$6.7}
\expandafter\def\csname val@rev.desc.d2.sspro.rep.haiku.C.D.rate01\endcsname{0.000}
\expandafter\def\csname cnt@rev.desc.d2.sspro.rep.haiku.C.D.rate01\endcsname{0}
\expandafter\def\csname pct@rev.desc.d2.sspro.rep.haiku.C.D.rate01\endcsname{0.0}
\expandafter\def\csname ppt@rev.desc.d2.sspro.rep.haiku.C.D.rate01\endcsname{0.0}
\expandafter\def\csname val@rev.desc.d2.sspro.rep.haiku.C.D.rate10\endcsname{0.067}
\expandafter\def\csname cnt@rev.desc.d2.sspro.rep.haiku.C.D.rate10\endcsname{0}
\expandafter\def\csname pct@rev.desc.d2.sspro.rep.haiku.C.D.rate10\endcsname{6.7}
\expandafter\def\csname ppt@rev.desc.d2.sspro.rep.haiku.C.D.rate10\endcsname{$+$6.7}
\expandafter\def\csname val@rev.desc.d2.sspro.rep.opus.C.D.acc-a\endcsname{0.643}
\expandafter\def\csname cnt@rev.desc.d2.sspro.rep.opus.C.D.acc-a\endcsname{1}
\expandafter\def\csname pct@rev.desc.d2.sspro.rep.opus.C.D.acc-a\endcsname{64.3}
\expandafter\def\csname ppt@rev.desc.d2.sspro.rep.opus.C.D.acc-a\endcsname{$+$64.3}
\expandafter\def\csname val@rev.desc.d2.sspro.rep.opus.C.D.acc-b\endcsname{0.643}
\expandafter\def\csname cnt@rev.desc.d2.sspro.rep.opus.C.D.acc-b\endcsname{1}
\expandafter\def\csname pct@rev.desc.d2.sspro.rep.opus.C.D.acc-b\endcsname{64.3}
\expandafter\def\csname ppt@rev.desc.d2.sspro.rep.opus.C.D.acc-b\endcsname{$+$64.3}
\expandafter\def\csname val@rev.desc.d2.sspro.rep.opus.C.D.cond01\endcsname{0.000}
\expandafter\def\csname cnt@rev.desc.d2.sspro.rep.opus.C.D.cond01\endcsname{0}
\expandafter\def\csname pct@rev.desc.d2.sspro.rep.opus.C.D.cond01\endcsname{0.0}
\expandafter\def\csname ppt@rev.desc.d2.sspro.rep.opus.C.D.cond01\endcsname{0.0}
\expandafter\def\csname val@rev.desc.d2.sspro.rep.opus.C.D.cond10\endcsname{0.000}
\expandafter\def\csname cnt@rev.desc.d2.sspro.rep.opus.C.D.cond10\endcsname{0}
\expandafter\def\csname pct@rev.desc.d2.sspro.rep.opus.C.D.cond10\endcsname{0.0}
\expandafter\def\csname ppt@rev.desc.d2.sspro.rep.opus.C.D.cond10\endcsname{0.0}
\expandafter\def\csname val@rev.desc.d2.sspro.rep.opus.C.D.n\endcsname{14}
\expandafter\def\csname cnt@rev.desc.d2.sspro.rep.opus.C.D.n\endcsname{14}
\expandafter\def\csname pct@rev.desc.d2.sspro.rep.opus.C.D.n\endcsname{1400.0}
\expandafter\def\csname ppt@rev.desc.d2.sspro.rep.opus.C.D.n\endcsname{$+$1400.0}
\expandafter\def\csname val@rev.desc.d2.sspro.rep.opus.C.D.n-apps\endcsname{6}
\expandafter\def\csname cnt@rev.desc.d2.sspro.rep.opus.C.D.n-apps\endcsname{6}
\expandafter\def\csname pct@rev.desc.d2.sspro.rep.opus.C.D.n-apps\endcsname{600.0}
\expandafter\def\csname ppt@rev.desc.d2.sspro.rep.opus.C.D.n-apps\endcsname{$+$600.0}
\expandafter\def\csname val@rev.desc.d2.sspro.rep.opus.C.D.n01\endcsname{0}
\expandafter\def\csname cnt@rev.desc.d2.sspro.rep.opus.C.D.n01\endcsname{0}
\expandafter\def\csname pct@rev.desc.d2.sspro.rep.opus.C.D.n01\endcsname{0.0}
\expandafter\def\csname ppt@rev.desc.d2.sspro.rep.opus.C.D.n01\endcsname{0.0}
\expandafter\def\csname val@rev.desc.d2.sspro.rep.opus.C.D.n10\endcsname{0}
\expandafter\def\csname cnt@rev.desc.d2.sspro.rep.opus.C.D.n10\endcsname{0}
\expandafter\def\csname pct@rev.desc.d2.sspro.rep.opus.C.D.n10\endcsname{0.0}
\expandafter\def\csname ppt@rev.desc.d2.sspro.rep.opus.C.D.n10\endcsname{0.0}
\expandafter\def\csname val@rev.desc.d2.sspro.rep.opus.C.D.rate-any\endcsname{0.000}
\expandafter\def\csname cnt@rev.desc.d2.sspro.rep.opus.C.D.rate-any\endcsname{0}
\expandafter\def\csname pct@rev.desc.d2.sspro.rep.opus.C.D.rate-any\endcsname{0.0}
\expandafter\def\csname ppt@rev.desc.d2.sspro.rep.opus.C.D.rate-any\endcsname{0.0}
\expandafter\def\csname val@rev.desc.d2.sspro.rep.opus.C.D.rate01\endcsname{0.000}
\expandafter\def\csname cnt@rev.desc.d2.sspro.rep.opus.C.D.rate01\endcsname{0}
\expandafter\def\csname pct@rev.desc.d2.sspro.rep.opus.C.D.rate01\endcsname{0.0}
\expandafter\def\csname ppt@rev.desc.d2.sspro.rep.opus.C.D.rate01\endcsname{0.0}
\expandafter\def\csname val@rev.desc.d2.sspro.rep.opus.C.D.rate10\endcsname{0.000}
\expandafter\def\csname cnt@rev.desc.d2.sspro.rep.opus.C.D.rate10\endcsname{0}
\expandafter\def\csname pct@rev.desc.d2.sspro.rep.opus.C.D.rate10\endcsname{0.0}
\expandafter\def\csname ppt@rev.desc.d2.sspro.rep.opus.C.D.rate10\endcsname{0.0}
\expandafter\def\csname val@rev.desc.d2.sspro.rep.opus.T.D.acc-a\endcsname{0.760}
\expandafter\def\csname cnt@rev.desc.d2.sspro.rep.opus.T.D.acc-a\endcsname{1}
\expandafter\def\csname pct@rev.desc.d2.sspro.rep.opus.T.D.acc-a\endcsname{76.0}
\expandafter\def\csname ppt@rev.desc.d2.sspro.rep.opus.T.D.acc-a\endcsname{$+$76.0}
\expandafter\def\csname val@rev.desc.d2.sspro.rep.opus.T.D.acc-b\endcsname{0.760}
\expandafter\def\csname cnt@rev.desc.d2.sspro.rep.opus.T.D.acc-b\endcsname{1}
\expandafter\def\csname pct@rev.desc.d2.sspro.rep.opus.T.D.acc-b\endcsname{76.0}
\expandafter\def\csname ppt@rev.desc.d2.sspro.rep.opus.T.D.acc-b\endcsname{$+$76.0}
\expandafter\def\csname val@rev.desc.d2.sspro.rep.opus.T.D.cond01\endcsname{0.217}
\expandafter\def\csname cnt@rev.desc.d2.sspro.rep.opus.T.D.cond01\endcsname{0}
\expandafter\def\csname pct@rev.desc.d2.sspro.rep.opus.T.D.cond01\endcsname{21.7}
\expandafter\def\csname ppt@rev.desc.d2.sspro.rep.opus.T.D.cond01\endcsname{$+$21.7}
\expandafter\def\csname val@rev.desc.d2.sspro.rep.opus.T.D.cond10\endcsname{0.068}
\expandafter\def\csname cnt@rev.desc.d2.sspro.rep.opus.T.D.cond10\endcsname{0}
\expandafter\def\csname pct@rev.desc.d2.sspro.rep.opus.T.D.cond10\endcsname{6.8}
\expandafter\def\csname ppt@rev.desc.d2.sspro.rep.opus.T.D.cond10\endcsname{$+$6.8}
\expandafter\def\csname val@rev.desc.d2.sspro.rep.opus.T.D.n\endcsname{96}
\expandafter\def\csname cnt@rev.desc.d2.sspro.rep.opus.T.D.n\endcsname{96}
\expandafter\def\csname pct@rev.desc.d2.sspro.rep.opus.T.D.n\endcsname{9600.0}
\expandafter\def\csname ppt@rev.desc.d2.sspro.rep.opus.T.D.n\endcsname{$+$9600.0}
\expandafter\def\csname val@rev.desc.d2.sspro.rep.opus.T.D.n-apps\endcsname{12}
\expandafter\def\csname cnt@rev.desc.d2.sspro.rep.opus.T.D.n-apps\endcsname{12}
\expandafter\def\csname pct@rev.desc.d2.sspro.rep.opus.T.D.n-apps\endcsname{1200.0}
\expandafter\def\csname ppt@rev.desc.d2.sspro.rep.opus.T.D.n-apps\endcsname{$+$1200.0}
\expandafter\def\csname val@rev.desc.d2.sspro.rep.opus.T.D.n01\endcsname{5}
\expandafter\def\csname cnt@rev.desc.d2.sspro.rep.opus.T.D.n01\endcsname{5}
\expandafter\def\csname pct@rev.desc.d2.sspro.rep.opus.T.D.n01\endcsname{500.0}
\expandafter\def\csname ppt@rev.desc.d2.sspro.rep.opus.T.D.n01\endcsname{$+$500.0}
\expandafter\def\csname val@rev.desc.d2.sspro.rep.opus.T.D.n10\endcsname{5}
\expandafter\def\csname cnt@rev.desc.d2.sspro.rep.opus.T.D.n10\endcsname{5}
\expandafter\def\csname pct@rev.desc.d2.sspro.rep.opus.T.D.n10\endcsname{500.0}
\expandafter\def\csname ppt@rev.desc.d2.sspro.rep.opus.T.D.n10\endcsname{$+$500.0}
\expandafter\def\csname val@rev.desc.d2.sspro.rep.opus.T.D.rate-any\endcsname{0.104}
\expandafter\def\csname cnt@rev.desc.d2.sspro.rep.opus.T.D.rate-any\endcsname{0}
\expandafter\def\csname pct@rev.desc.d2.sspro.rep.opus.T.D.rate-any\endcsname{10.4}
\expandafter\def\csname ppt@rev.desc.d2.sspro.rep.opus.T.D.rate-any\endcsname{$+$10.4}
\expandafter\def\csname val@rev.desc.d2.sspro.rep.opus.T.D.rate01\endcsname{0.052}
\expandafter\def\csname cnt@rev.desc.d2.sspro.rep.opus.T.D.rate01\endcsname{0}
\expandafter\def\csname pct@rev.desc.d2.sspro.rep.opus.T.D.rate01\endcsname{5.2}
\expandafter\def\csname ppt@rev.desc.d2.sspro.rep.opus.T.D.rate01\endcsname{$+$5.2}
\expandafter\def\csname val@rev.desc.d2.sspro.rep.opus.T.D.rate10\endcsname{0.052}
\expandafter\def\csname cnt@rev.desc.d2.sspro.rep.opus.T.D.rate10\endcsname{0}
\expandafter\def\csname pct@rev.desc.d2.sspro.rep.opus.T.D.rate10\endcsname{5.2}
\expandafter\def\csname ppt@rev.desc.d2.sspro.rep.opus.T.D.rate10\endcsname{$+$5.2}
\expandafter\def\csname val@rev.desc.d2.sspro.rep.sonnet.C.D.acc-a\endcsname{0.500}
\expandafter\def\csname cnt@rev.desc.d2.sspro.rep.sonnet.C.D.acc-a\endcsname{0}
\expandafter\def\csname pct@rev.desc.d2.sspro.rep.sonnet.C.D.acc-a\endcsname{50.0}
\expandafter\def\csname ppt@rev.desc.d2.sspro.rep.sonnet.C.D.acc-a\endcsname{$+$50.0}
\expandafter\def\csname val@rev.desc.d2.sspro.rep.sonnet.C.D.acc-b\endcsname{0.625}
\expandafter\def\csname cnt@rev.desc.d2.sspro.rep.sonnet.C.D.acc-b\endcsname{1}
\expandafter\def\csname pct@rev.desc.d2.sspro.rep.sonnet.C.D.acc-b\endcsname{62.5}
\expandafter\def\csname ppt@rev.desc.d2.sspro.rep.sonnet.C.D.acc-b\endcsname{$+$62.5}
\expandafter\def\csname val@rev.desc.d2.sspro.rep.sonnet.C.D.cond01\endcsname{0.250}
\expandafter\def\csname cnt@rev.desc.d2.sspro.rep.sonnet.C.D.cond01\endcsname{0}
\expandafter\def\csname pct@rev.desc.d2.sspro.rep.sonnet.C.D.cond01\endcsname{25.0}
\expandafter\def\csname ppt@rev.desc.d2.sspro.rep.sonnet.C.D.cond01\endcsname{$+$25.0}
\expandafter\def\csname val@rev.desc.d2.sspro.rep.sonnet.C.D.cond10\endcsname{0.000}
\expandafter\def\csname cnt@rev.desc.d2.sspro.rep.sonnet.C.D.cond10\endcsname{0}
\expandafter\def\csname pct@rev.desc.d2.sspro.rep.sonnet.C.D.cond10\endcsname{0.0}
\expandafter\def\csname ppt@rev.desc.d2.sspro.rep.sonnet.C.D.cond10\endcsname{0.0}
\expandafter\def\csname val@rev.desc.d2.sspro.rep.sonnet.C.D.n\endcsname{16}
\expandafter\def\csname cnt@rev.desc.d2.sspro.rep.sonnet.C.D.n\endcsname{16}
\expandafter\def\csname pct@rev.desc.d2.sspro.rep.sonnet.C.D.n\endcsname{1600.0}
\expandafter\def\csname ppt@rev.desc.d2.sspro.rep.sonnet.C.D.n\endcsname{$+$1600.0}
\expandafter\def\csname val@rev.desc.d2.sspro.rep.sonnet.C.D.n-apps\endcsname{6}
\expandafter\def\csname cnt@rev.desc.d2.sspro.rep.sonnet.C.D.n-apps\endcsname{6}
\expandafter\def\csname pct@rev.desc.d2.sspro.rep.sonnet.C.D.n-apps\endcsname{600.0}
\expandafter\def\csname ppt@rev.desc.d2.sspro.rep.sonnet.C.D.n-apps\endcsname{$+$600.0}
\expandafter\def\csname val@rev.desc.d2.sspro.rep.sonnet.C.D.n01\endcsname{2}
\expandafter\def\csname cnt@rev.desc.d2.sspro.rep.sonnet.C.D.n01\endcsname{2}
\expandafter\def\csname pct@rev.desc.d2.sspro.rep.sonnet.C.D.n01\endcsname{200.0}
\expandafter\def\csname ppt@rev.desc.d2.sspro.rep.sonnet.C.D.n01\endcsname{$+$200.0}
\expandafter\def\csname val@rev.desc.d2.sspro.rep.sonnet.C.D.n10\endcsname{0}
\expandafter\def\csname cnt@rev.desc.d2.sspro.rep.sonnet.C.D.n10\endcsname{0}
\expandafter\def\csname pct@rev.desc.d2.sspro.rep.sonnet.C.D.n10\endcsname{0.0}
\expandafter\def\csname ppt@rev.desc.d2.sspro.rep.sonnet.C.D.n10\endcsname{0.0}
\expandafter\def\csname val@rev.desc.d2.sspro.rep.sonnet.C.D.rate-any\endcsname{0.125}
\expandafter\def\csname cnt@rev.desc.d2.sspro.rep.sonnet.C.D.rate-any\endcsname{0}
\expandafter\def\csname pct@rev.desc.d2.sspro.rep.sonnet.C.D.rate-any\endcsname{12.5}
\expandafter\def\csname ppt@rev.desc.d2.sspro.rep.sonnet.C.D.rate-any\endcsname{$+$12.5}
\expandafter\def\csname val@rev.desc.d2.sspro.rep.sonnet.C.D.rate01\endcsname{0.125}
\expandafter\def\csname cnt@rev.desc.d2.sspro.rep.sonnet.C.D.rate01\endcsname{0}
\expandafter\def\csname pct@rev.desc.d2.sspro.rep.sonnet.C.D.rate01\endcsname{12.5}
\expandafter\def\csname ppt@rev.desc.d2.sspro.rep.sonnet.C.D.rate01\endcsname{$+$12.5}
\expandafter\def\csname val@rev.desc.d2.sspro.rep.sonnet.C.D.rate10\endcsname{0.000}
\expandafter\def\csname cnt@rev.desc.d2.sspro.rep.sonnet.C.D.rate10\endcsname{0}
\expandafter\def\csname pct@rev.desc.d2.sspro.rep.sonnet.C.D.rate10\endcsname{0.0}
\expandafter\def\csname ppt@rev.desc.d2.sspro.rep.sonnet.C.D.rate10\endcsname{0.0}
\expandafter\def\csname val@rev.desc.d2.sspro.rep.sonnet.T.D.acc-a\endcsname{0.740}
\expandafter\def\csname cnt@rev.desc.d2.sspro.rep.sonnet.T.D.acc-a\endcsname{1}
\expandafter\def\csname pct@rev.desc.d2.sspro.rep.sonnet.T.D.acc-a\endcsname{74.0}
\expandafter\def\csname ppt@rev.desc.d2.sspro.rep.sonnet.T.D.acc-a\endcsname{$+$74.0}
\expandafter\def\csname val@rev.desc.d2.sspro.rep.sonnet.T.D.acc-b\endcsname{0.708}
\expandafter\def\csname cnt@rev.desc.d2.sspro.rep.sonnet.T.D.acc-b\endcsname{1}
\expandafter\def\csname pct@rev.desc.d2.sspro.rep.sonnet.T.D.acc-b\endcsname{70.8}
\expandafter\def\csname ppt@rev.desc.d2.sspro.rep.sonnet.T.D.acc-b\endcsname{$+$70.8}
\expandafter\def\csname val@rev.desc.d2.sspro.rep.sonnet.T.D.cond01\endcsname{0.120}
\expandafter\def\csname cnt@rev.desc.d2.sspro.rep.sonnet.T.D.cond01\endcsname{0}
\expandafter\def\csname pct@rev.desc.d2.sspro.rep.sonnet.T.D.cond01\endcsname{12.0}
\expandafter\def\csname ppt@rev.desc.d2.sspro.rep.sonnet.T.D.cond01\endcsname{$+$12.0}
\expandafter\def\csname val@rev.desc.d2.sspro.rep.sonnet.T.D.cond10\endcsname{0.085}
\expandafter\def\csname cnt@rev.desc.d2.sspro.rep.sonnet.T.D.cond10\endcsname{0}
\expandafter\def\csname pct@rev.desc.d2.sspro.rep.sonnet.T.D.cond10\endcsname{8.5}
\expandafter\def\csname ppt@rev.desc.d2.sspro.rep.sonnet.T.D.cond10\endcsname{$+$8.5}
\expandafter\def\csname val@rev.desc.d2.sspro.rep.sonnet.T.D.n\endcsname{96}
\expandafter\def\csname cnt@rev.desc.d2.sspro.rep.sonnet.T.D.n\endcsname{96}
\expandafter\def\csname pct@rev.desc.d2.sspro.rep.sonnet.T.D.n\endcsname{9600.0}
\expandafter\def\csname ppt@rev.desc.d2.sspro.rep.sonnet.T.D.n\endcsname{$+$9600.0}
\expandafter\def\csname val@rev.desc.d2.sspro.rep.sonnet.T.D.n-apps\endcsname{12}
\expandafter\def\csname cnt@rev.desc.d2.sspro.rep.sonnet.T.D.n-apps\endcsname{12}
\expandafter\def\csname pct@rev.desc.d2.sspro.rep.sonnet.T.D.n-apps\endcsname{1200.0}
\expandafter\def\csname ppt@rev.desc.d2.sspro.rep.sonnet.T.D.n-apps\endcsname{$+$1200.0}
\expandafter\def\csname val@rev.desc.d2.sspro.rep.sonnet.T.D.n01\endcsname{3}
\expandafter\def\csname cnt@rev.desc.d2.sspro.rep.sonnet.T.D.n01\endcsname{3}
\expandafter\def\csname pct@rev.desc.d2.sspro.rep.sonnet.T.D.n01\endcsname{300.0}
\expandafter\def\csname ppt@rev.desc.d2.sspro.rep.sonnet.T.D.n01\endcsname{$+$300.0}
\expandafter\def\csname val@rev.desc.d2.sspro.rep.sonnet.T.D.n10\endcsname{6}
\expandafter\def\csname cnt@rev.desc.d2.sspro.rep.sonnet.T.D.n10\endcsname{6}
\expandafter\def\csname pct@rev.desc.d2.sspro.rep.sonnet.T.D.n10\endcsname{600.0}
\expandafter\def\csname ppt@rev.desc.d2.sspro.rep.sonnet.T.D.n10\endcsname{$+$600.0}
\expandafter\def\csname val@rev.desc.d2.sspro.rep.sonnet.T.D.rate-any\endcsname{0.094}
\expandafter\def\csname cnt@rev.desc.d2.sspro.rep.sonnet.T.D.rate-any\endcsname{0}
\expandafter\def\csname pct@rev.desc.d2.sspro.rep.sonnet.T.D.rate-any\endcsname{9.4}
\expandafter\def\csname ppt@rev.desc.d2.sspro.rep.sonnet.T.D.rate-any\endcsname{$+$9.4}
\expandafter\def\csname val@rev.desc.d2.sspro.rep.sonnet.T.D.rate01\endcsname{0.031}
\expandafter\def\csname cnt@rev.desc.d2.sspro.rep.sonnet.T.D.rate01\endcsname{0}
\expandafter\def\csname pct@rev.desc.d2.sspro.rep.sonnet.T.D.rate01\endcsname{3.1}
\expandafter\def\csname ppt@rev.desc.d2.sspro.rep.sonnet.T.D.rate01\endcsname{$+$3.1}
\expandafter\def\csname val@rev.desc.d2.sspro.rep.sonnet.T.D.rate10\endcsname{0.062}
\expandafter\def\csname cnt@rev.desc.d2.sspro.rep.sonnet.T.D.rate10\endcsname{0}
\expandafter\def\csname pct@rev.desc.d2.sspro.rep.sonnet.T.D.rate10\endcsname{6.2}
\expandafter\def\csname ppt@rev.desc.d2.sspro.rep.sonnet.T.D.rate10\endcsname{$+$6.2}
\expandafter\def\csname val@rev.desc.d2.uiv.p1.C.D.acc-a\endcsname{0.635}
\expandafter\def\csname cnt@rev.desc.d2.uiv.p1.C.D.acc-a\endcsname{1}
\expandafter\def\csname pct@rev.desc.d2.uiv.p1.C.D.acc-a\endcsname{63.5}
\expandafter\def\csname ppt@rev.desc.d2.uiv.p1.C.D.acc-a\endcsname{$+$63.5}
\expandafter\def\csname val@rev.desc.d2.uiv.p1.C.D.acc-b\endcsname{0.760}
\expandafter\def\csname cnt@rev.desc.d2.uiv.p1.C.D.acc-b\endcsname{1}
\expandafter\def\csname pct@rev.desc.d2.uiv.p1.C.D.acc-b\endcsname{76.0}
\expandafter\def\csname ppt@rev.desc.d2.uiv.p1.C.D.acc-b\endcsname{$+$76.0}
\expandafter\def\csname val@rev.desc.d2.uiv.p1.C.D.cond01\endcsname{0.457}
\expandafter\def\csname cnt@rev.desc.d2.uiv.p1.C.D.cond01\endcsname{0}
\expandafter\def\csname pct@rev.desc.d2.uiv.p1.C.D.cond01\endcsname{45.7}
\expandafter\def\csname ppt@rev.desc.d2.uiv.p1.C.D.cond01\endcsname{$+$45.7}
\expandafter\def\csname val@rev.desc.d2.uiv.p1.C.D.cond10\endcsname{0.066}
\expandafter\def\csname cnt@rev.desc.d2.uiv.p1.C.D.cond10\endcsname{0}
\expandafter\def\csname pct@rev.desc.d2.uiv.p1.C.D.cond10\endcsname{6.6}
\expandafter\def\csname ppt@rev.desc.d2.uiv.p1.C.D.cond10\endcsname{$+$6.6}
\expandafter\def\csname val@rev.desc.d2.uiv.p1.C.D.n\endcsname{96}
\expandafter\def\csname cnt@rev.desc.d2.uiv.p1.C.D.n\endcsname{96}
\expandafter\def\csname pct@rev.desc.d2.uiv.p1.C.D.n\endcsname{9600.0}
\expandafter\def\csname ppt@rev.desc.d2.uiv.p1.C.D.n\endcsname{$+$9600.0}
\expandafter\def\csname val@rev.desc.d2.uiv.p1.C.D.n-apps\endcsname{12}
\expandafter\def\csname cnt@rev.desc.d2.uiv.p1.C.D.n-apps\endcsname{12}
\expandafter\def\csname pct@rev.desc.d2.uiv.p1.C.D.n-apps\endcsname{1200.0}
\expandafter\def\csname ppt@rev.desc.d2.uiv.p1.C.D.n-apps\endcsname{$+$1200.0}
\expandafter\def\csname val@rev.desc.d2.uiv.p1.C.D.n01\endcsname{16}
\expandafter\def\csname cnt@rev.desc.d2.uiv.p1.C.D.n01\endcsname{16}
\expandafter\def\csname pct@rev.desc.d2.uiv.p1.C.D.n01\endcsname{1600.0}
\expandafter\def\csname ppt@rev.desc.d2.uiv.p1.C.D.n01\endcsname{$+$1600.0}
\expandafter\def\csname val@rev.desc.d2.uiv.p1.C.D.n10\endcsname{4}
\expandafter\def\csname cnt@rev.desc.d2.uiv.p1.C.D.n10\endcsname{4}
\expandafter\def\csname pct@rev.desc.d2.uiv.p1.C.D.n10\endcsname{400.0}
\expandafter\def\csname ppt@rev.desc.d2.uiv.p1.C.D.n10\endcsname{$+$400.0}
\expandafter\def\csname val@rev.desc.d2.uiv.p1.C.D.rate-any\endcsname{0.208}
\expandafter\def\csname cnt@rev.desc.d2.uiv.p1.C.D.rate-any\endcsname{0}
\expandafter\def\csname pct@rev.desc.d2.uiv.p1.C.D.rate-any\endcsname{20.8}
\expandafter\def\csname ppt@rev.desc.d2.uiv.p1.C.D.rate-any\endcsname{$+$20.8}
\expandafter\def\csname val@rev.desc.d2.uiv.p1.C.D.rate01\endcsname{0.167}
\expandafter\def\csname cnt@rev.desc.d2.uiv.p1.C.D.rate01\endcsname{0}
\expandafter\def\csname pct@rev.desc.d2.uiv.p1.C.D.rate01\endcsname{16.7}
\expandafter\def\csname ppt@rev.desc.d2.uiv.p1.C.D.rate01\endcsname{$+$16.7}
\expandafter\def\csname val@rev.desc.d2.uiv.p1.C.D.rate10\endcsname{0.042}
\expandafter\def\csname cnt@rev.desc.d2.uiv.p1.C.D.rate10\endcsname{0}
\expandafter\def\csname pct@rev.desc.d2.uiv.p1.C.D.rate10\endcsname{4.2}
\expandafter\def\csname ppt@rev.desc.d2.uiv.p1.C.D.rate10\endcsname{$+$4.2}
\expandafter\def\csname val@rev.desc.d2.uiv.p1.C.F.center-d1.acc-a\endcsname{0.562}
\expandafter\def\csname cnt@rev.desc.d2.uiv.p1.C.F.center-d1.acc-a\endcsname{1}
\expandafter\def\csname pct@rev.desc.d2.uiv.p1.C.F.center-d1.acc-a\endcsname{56.2}
\expandafter\def\csname ppt@rev.desc.d2.uiv.p1.C.F.center-d1.acc-a\endcsname{$+$56.2}
\expandafter\def\csname val@rev.desc.d2.uiv.p1.C.F.center-d1.acc-b\endcsname{0.771}
\expandafter\def\csname cnt@rev.desc.d2.uiv.p1.C.F.center-d1.acc-b\endcsname{1}
\expandafter\def\csname pct@rev.desc.d2.uiv.p1.C.F.center-d1.acc-b\endcsname{77.1}
\expandafter\def\csname ppt@rev.desc.d2.uiv.p1.C.F.center-d1.acc-b\endcsname{$+$77.1}
\expandafter\def\csname val@rev.desc.d2.uiv.p1.C.F.center-d1.cond01\endcsname{0.500}
\expandafter\def\csname cnt@rev.desc.d2.uiv.p1.C.F.center-d1.cond01\endcsname{0}
\expandafter\def\csname pct@rev.desc.d2.uiv.p1.C.F.center-d1.cond01\endcsname{50.0}
\expandafter\def\csname ppt@rev.desc.d2.uiv.p1.C.F.center-d1.cond01\endcsname{$+$50.0}
\expandafter\def\csname val@rev.desc.d2.uiv.p1.C.F.center-d1.cond10\endcsname{0.019}
\expandafter\def\csname cnt@rev.desc.d2.uiv.p1.C.F.center-d1.cond10\endcsname{0}
\expandafter\def\csname pct@rev.desc.d2.uiv.p1.C.F.center-d1.cond10\endcsname{1.9}
\expandafter\def\csname ppt@rev.desc.d2.uiv.p1.C.F.center-d1.cond10\endcsname{$+$1.9}
\expandafter\def\csname val@rev.desc.d2.uiv.p1.C.F.center-d1.n\endcsname{96}
\expandafter\def\csname cnt@rev.desc.d2.uiv.p1.C.F.center-d1.n\endcsname{96}
\expandafter\def\csname pct@rev.desc.d2.uiv.p1.C.F.center-d1.n\endcsname{9600.0}
\expandafter\def\csname ppt@rev.desc.d2.uiv.p1.C.F.center-d1.n\endcsname{$+$9600.0}
\expandafter\def\csname val@rev.desc.d2.uiv.p1.C.F.center-d1.n-apps\endcsname{12}
\expandafter\def\csname cnt@rev.desc.d2.uiv.p1.C.F.center-d1.n-apps\endcsname{12}
\expandafter\def\csname pct@rev.desc.d2.uiv.p1.C.F.center-d1.n-apps\endcsname{1200.0}
\expandafter\def\csname ppt@rev.desc.d2.uiv.p1.C.F.center-d1.n-apps\endcsname{$+$1200.0}
\expandafter\def\csname val@rev.desc.d2.uiv.p1.C.F.center-d1.n01\endcsname{21}
\expandafter\def\csname cnt@rev.desc.d2.uiv.p1.C.F.center-d1.n01\endcsname{21}
\expandafter\def\csname pct@rev.desc.d2.uiv.p1.C.F.center-d1.n01\endcsname{2100.0}
\expandafter\def\csname ppt@rev.desc.d2.uiv.p1.C.F.center-d1.n01\endcsname{$+$2100.0}
\expandafter\def\csname val@rev.desc.d2.uiv.p1.C.F.center-d1.n10\endcsname{1}
\expandafter\def\csname cnt@rev.desc.d2.uiv.p1.C.F.center-d1.n10\endcsname{1}
\expandafter\def\csname pct@rev.desc.d2.uiv.p1.C.F.center-d1.n10\endcsname{100.0}
\expandafter\def\csname ppt@rev.desc.d2.uiv.p1.C.F.center-d1.n10\endcsname{$+$100.0}
\expandafter\def\csname val@rev.desc.d2.uiv.p1.C.F.center-d1.rate-any\endcsname{0.229}
\expandafter\def\csname cnt@rev.desc.d2.uiv.p1.C.F.center-d1.rate-any\endcsname{0}
\expandafter\def\csname pct@rev.desc.d2.uiv.p1.C.F.center-d1.rate-any\endcsname{22.9}
\expandafter\def\csname ppt@rev.desc.d2.uiv.p1.C.F.center-d1.rate-any\endcsname{$+$22.9}
\expandafter\def\csname val@rev.desc.d2.uiv.p1.C.F.center-d1.rate01\endcsname{0.219}
\expandafter\def\csname cnt@rev.desc.d2.uiv.p1.C.F.center-d1.rate01\endcsname{0}
\expandafter\def\csname pct@rev.desc.d2.uiv.p1.C.F.center-d1.rate01\endcsname{21.9}
\expandafter\def\csname ppt@rev.desc.d2.uiv.p1.C.F.center-d1.rate01\endcsname{$+$21.9}
\expandafter\def\csname val@rev.desc.d2.uiv.p1.C.F.center-d1.rate10\endcsname{0.010}
\expandafter\def\csname cnt@rev.desc.d2.uiv.p1.C.F.center-d1.rate10\endcsname{0}
\expandafter\def\csname pct@rev.desc.d2.uiv.p1.C.F.center-d1.rate10\endcsname{1.0}
\expandafter\def\csname ppt@rev.desc.d2.uiv.p1.C.F.center-d1.rate10\endcsname{$+$1.0}
\expandafter\def\csname val@rev.desc.d2.uiv.p1.C.F.context-d1.acc-a\endcsname{0.562}
\expandafter\def\csname cnt@rev.desc.d2.uiv.p1.C.F.context-d1.acc-a\endcsname{1}
\expandafter\def\csname pct@rev.desc.d2.uiv.p1.C.F.context-d1.acc-a\endcsname{56.2}
\expandafter\def\csname ppt@rev.desc.d2.uiv.p1.C.F.context-d1.acc-a\endcsname{$+$56.2}
\expandafter\def\csname val@rev.desc.d2.uiv.p1.C.F.context-d1.acc-b\endcsname{0.740}
\expandafter\def\csname cnt@rev.desc.d2.uiv.p1.C.F.context-d1.acc-b\endcsname{1}
\expandafter\def\csname pct@rev.desc.d2.uiv.p1.C.F.context-d1.acc-b\endcsname{74.0}
\expandafter\def\csname ppt@rev.desc.d2.uiv.p1.C.F.context-d1.acc-b\endcsname{$+$74.0}
\expandafter\def\csname val@rev.desc.d2.uiv.p1.C.F.context-d1.cond01\endcsname{0.452}
\expandafter\def\csname cnt@rev.desc.d2.uiv.p1.C.F.context-d1.cond01\endcsname{0}
\expandafter\def\csname pct@rev.desc.d2.uiv.p1.C.F.context-d1.cond01\endcsname{45.2}
\expandafter\def\csname ppt@rev.desc.d2.uiv.p1.C.F.context-d1.cond01\endcsname{$+$45.2}
\expandafter\def\csname val@rev.desc.d2.uiv.p1.C.F.context-d1.cond10\endcsname{0.037}
\expandafter\def\csname cnt@rev.desc.d2.uiv.p1.C.F.context-d1.cond10\endcsname{0}
\expandafter\def\csname pct@rev.desc.d2.uiv.p1.C.F.context-d1.cond10\endcsname{3.7}
\expandafter\def\csname ppt@rev.desc.d2.uiv.p1.C.F.context-d1.cond10\endcsname{$+$3.7}
\expandafter\def\csname val@rev.desc.d2.uiv.p1.C.F.context-d1.n\endcsname{96}
\expandafter\def\csname cnt@rev.desc.d2.uiv.p1.C.F.context-d1.n\endcsname{96}
\expandafter\def\csname pct@rev.desc.d2.uiv.p1.C.F.context-d1.n\endcsname{9600.0}
\expandafter\def\csname ppt@rev.desc.d2.uiv.p1.C.F.context-d1.n\endcsname{$+$9600.0}
\expandafter\def\csname val@rev.desc.d2.uiv.p1.C.F.context-d1.n-apps\endcsname{12}
\expandafter\def\csname cnt@rev.desc.d2.uiv.p1.C.F.context-d1.n-apps\endcsname{12}
\expandafter\def\csname pct@rev.desc.d2.uiv.p1.C.F.context-d1.n-apps\endcsname{1200.0}
\expandafter\def\csname ppt@rev.desc.d2.uiv.p1.C.F.context-d1.n-apps\endcsname{$+$1200.0}
\expandafter\def\csname val@rev.desc.d2.uiv.p1.C.F.context-d1.n01\endcsname{19}
\expandafter\def\csname cnt@rev.desc.d2.uiv.p1.C.F.context-d1.n01\endcsname{19}
\expandafter\def\csname pct@rev.desc.d2.uiv.p1.C.F.context-d1.n01\endcsname{1900.0}
\expandafter\def\csname ppt@rev.desc.d2.uiv.p1.C.F.context-d1.n01\endcsname{$+$1900.0}
\expandafter\def\csname val@rev.desc.d2.uiv.p1.C.F.context-d1.n10\endcsname{2}
\expandafter\def\csname cnt@rev.desc.d2.uiv.p1.C.F.context-d1.n10\endcsname{2}
\expandafter\def\csname pct@rev.desc.d2.uiv.p1.C.F.context-d1.n10\endcsname{200.0}
\expandafter\def\csname ppt@rev.desc.d2.uiv.p1.C.F.context-d1.n10\endcsname{$+$200.0}
\expandafter\def\csname val@rev.desc.d2.uiv.p1.C.F.context-d1.rate-any\endcsname{0.219}
\expandafter\def\csname cnt@rev.desc.d2.uiv.p1.C.F.context-d1.rate-any\endcsname{0}
\expandafter\def\csname pct@rev.desc.d2.uiv.p1.C.F.context-d1.rate-any\endcsname{21.9}
\expandafter\def\csname ppt@rev.desc.d2.uiv.p1.C.F.context-d1.rate-any\endcsname{$+$21.9}
\expandafter\def\csname val@rev.desc.d2.uiv.p1.C.F.context-d1.rate01\endcsname{0.198}
\expandafter\def\csname cnt@rev.desc.d2.uiv.p1.C.F.context-d1.rate01\endcsname{0}
\expandafter\def\csname pct@rev.desc.d2.uiv.p1.C.F.context-d1.rate01\endcsname{19.8}
\expandafter\def\csname ppt@rev.desc.d2.uiv.p1.C.F.context-d1.rate01\endcsname{$+$19.8}
\expandafter\def\csname val@rev.desc.d2.uiv.p1.C.F.context-d1.rate10\endcsname{0.021}
\expandafter\def\csname cnt@rev.desc.d2.uiv.p1.C.F.context-d1.rate10\endcsname{0}
\expandafter\def\csname pct@rev.desc.d2.uiv.p1.C.F.context-d1.rate10\endcsname{2.1}
\expandafter\def\csname ppt@rev.desc.d2.uiv.p1.C.F.context-d1.rate10\endcsname{$+$2.1}
\expandafter\def\csname val@rev.desc.d2.uiv.p1.C.F.dual-d2.acc-a\endcsname{0.552}
\expandafter\def\csname cnt@rev.desc.d2.uiv.p1.C.F.dual-d2.acc-a\endcsname{1}
\expandafter\def\csname pct@rev.desc.d2.uiv.p1.C.F.dual-d2.acc-a\endcsname{55.2}
\expandafter\def\csname ppt@rev.desc.d2.uiv.p1.C.F.dual-d2.acc-a\endcsname{$+$55.2}
\expandafter\def\csname val@rev.desc.d2.uiv.p1.C.F.dual-d2.acc-b\endcsname{0.760}
\expandafter\def\csname cnt@rev.desc.d2.uiv.p1.C.F.dual-d2.acc-b\endcsname{1}
\expandafter\def\csname pct@rev.desc.d2.uiv.p1.C.F.dual-d2.acc-b\endcsname{76.0}
\expandafter\def\csname ppt@rev.desc.d2.uiv.p1.C.F.dual-d2.acc-b\endcsname{$+$76.0}
\expandafter\def\csname val@rev.desc.d2.uiv.p1.C.F.dual-d2.cond01\endcsname{0.512}
\expandafter\def\csname cnt@rev.desc.d2.uiv.p1.C.F.dual-d2.cond01\endcsname{1}
\expandafter\def\csname pct@rev.desc.d2.uiv.p1.C.F.dual-d2.cond01\endcsname{51.2}
\expandafter\def\csname ppt@rev.desc.d2.uiv.p1.C.F.dual-d2.cond01\endcsname{$+$51.2}
\expandafter\def\csname val@rev.desc.d2.uiv.p1.C.F.dual-d2.cond10\endcsname{0.038}
\expandafter\def\csname cnt@rev.desc.d2.uiv.p1.C.F.dual-d2.cond10\endcsname{0}
\expandafter\def\csname pct@rev.desc.d2.uiv.p1.C.F.dual-d2.cond10\endcsname{3.8}
\expandafter\def\csname ppt@rev.desc.d2.uiv.p1.C.F.dual-d2.cond10\endcsname{$+$3.8}
\expandafter\def\csname val@rev.desc.d2.uiv.p1.C.F.dual-d2.n\endcsname{96}
\expandafter\def\csname cnt@rev.desc.d2.uiv.p1.C.F.dual-d2.n\endcsname{96}
\expandafter\def\csname pct@rev.desc.d2.uiv.p1.C.F.dual-d2.n\endcsname{9600.0}
\expandafter\def\csname ppt@rev.desc.d2.uiv.p1.C.F.dual-d2.n\endcsname{$+$9600.0}
\expandafter\def\csname val@rev.desc.d2.uiv.p1.C.F.dual-d2.n-apps\endcsname{12}
\expandafter\def\csname cnt@rev.desc.d2.uiv.p1.C.F.dual-d2.n-apps\endcsname{12}
\expandafter\def\csname pct@rev.desc.d2.uiv.p1.C.F.dual-d2.n-apps\endcsname{1200.0}
\expandafter\def\csname ppt@rev.desc.d2.uiv.p1.C.F.dual-d2.n-apps\endcsname{$+$1200.0}
\expandafter\def\csname val@rev.desc.d2.uiv.p1.C.F.dual-d2.n01\endcsname{22}
\expandafter\def\csname cnt@rev.desc.d2.uiv.p1.C.F.dual-d2.n01\endcsname{22}
\expandafter\def\csname pct@rev.desc.d2.uiv.p1.C.F.dual-d2.n01\endcsname{2200.0}
\expandafter\def\csname ppt@rev.desc.d2.uiv.p1.C.F.dual-d2.n01\endcsname{$+$2200.0}
\expandafter\def\csname val@rev.desc.d2.uiv.p1.C.F.dual-d2.n10\endcsname{2}
\expandafter\def\csname cnt@rev.desc.d2.uiv.p1.C.F.dual-d2.n10\endcsname{2}
\expandafter\def\csname pct@rev.desc.d2.uiv.p1.C.F.dual-d2.n10\endcsname{200.0}
\expandafter\def\csname ppt@rev.desc.d2.uiv.p1.C.F.dual-d2.n10\endcsname{$+$200.0}
\expandafter\def\csname val@rev.desc.d2.uiv.p1.C.F.dual-d2.rate-any\endcsname{0.250}
\expandafter\def\csname cnt@rev.desc.d2.uiv.p1.C.F.dual-d2.rate-any\endcsname{0}
\expandafter\def\csname pct@rev.desc.d2.uiv.p1.C.F.dual-d2.rate-any\endcsname{25.0}
\expandafter\def\csname ppt@rev.desc.d2.uiv.p1.C.F.dual-d2.rate-any\endcsname{$+$25.0}
\expandafter\def\csname val@rev.desc.d2.uiv.p1.C.F.dual-d2.rate01\endcsname{0.229}
\expandafter\def\csname cnt@rev.desc.d2.uiv.p1.C.F.dual-d2.rate01\endcsname{0}
\expandafter\def\csname pct@rev.desc.d2.uiv.p1.C.F.dual-d2.rate01\endcsname{22.9}
\expandafter\def\csname ppt@rev.desc.d2.uiv.p1.C.F.dual-d2.rate01\endcsname{$+$22.9}
\expandafter\def\csname val@rev.desc.d2.uiv.p1.C.F.dual-d2.rate10\endcsname{0.021}
\expandafter\def\csname cnt@rev.desc.d2.uiv.p1.C.F.dual-d2.rate10\endcsname{0}
\expandafter\def\csname pct@rev.desc.d2.uiv.p1.C.F.dual-d2.rate10\endcsname{2.1}
\expandafter\def\csname ppt@rev.desc.d2.uiv.p1.C.F.dual-d2.rate10\endcsname{$+$2.1}
\expandafter\def\csname val@rev.desc.d2.uiv.p1.C.F.grid2-d3.acc-a\endcsname{0.490}
\expandafter\def\csname cnt@rev.desc.d2.uiv.p1.C.F.grid2-d3.acc-a\endcsname{0}
\expandafter\def\csname pct@rev.desc.d2.uiv.p1.C.F.grid2-d3.acc-a\endcsname{49.0}
\expandafter\def\csname ppt@rev.desc.d2.uiv.p1.C.F.grid2-d3.acc-a\endcsname{$+$49.0}
\expandafter\def\csname val@rev.desc.d2.uiv.p1.C.F.grid2-d3.acc-b\endcsname{0.729}
\expandafter\def\csname cnt@rev.desc.d2.uiv.p1.C.F.grid2-d3.acc-b\endcsname{1}
\expandafter\def\csname pct@rev.desc.d2.uiv.p1.C.F.grid2-d3.acc-b\endcsname{72.9}
\expandafter\def\csname ppt@rev.desc.d2.uiv.p1.C.F.grid2-d3.acc-b\endcsname{$+$72.9}
\expandafter\def\csname val@rev.desc.d2.uiv.p1.C.F.grid2-d3.cond01\endcsname{0.510}
\expandafter\def\csname cnt@rev.desc.d2.uiv.p1.C.F.grid2-d3.cond01\endcsname{1}
\expandafter\def\csname pct@rev.desc.d2.uiv.p1.C.F.grid2-d3.cond01\endcsname{51.0}
\expandafter\def\csname ppt@rev.desc.d2.uiv.p1.C.F.grid2-d3.cond01\endcsname{$+$51.0}
\expandafter\def\csname val@rev.desc.d2.uiv.p1.C.F.grid2-d3.cond10\endcsname{0.043}
\expandafter\def\csname cnt@rev.desc.d2.uiv.p1.C.F.grid2-d3.cond10\endcsname{0}
\expandafter\def\csname pct@rev.desc.d2.uiv.p1.C.F.grid2-d3.cond10\endcsname{4.3}
\expandafter\def\csname ppt@rev.desc.d2.uiv.p1.C.F.grid2-d3.cond10\endcsname{$+$4.3}
\expandafter\def\csname val@rev.desc.d2.uiv.p1.C.F.grid2-d3.n\endcsname{96}
\expandafter\def\csname cnt@rev.desc.d2.uiv.p1.C.F.grid2-d3.n\endcsname{96}
\expandafter\def\csname pct@rev.desc.d2.uiv.p1.C.F.grid2-d3.n\endcsname{9600.0}
\expandafter\def\csname ppt@rev.desc.d2.uiv.p1.C.F.grid2-d3.n\endcsname{$+$9600.0}
\expandafter\def\csname val@rev.desc.d2.uiv.p1.C.F.grid2-d3.n-apps\endcsname{12}
\expandafter\def\csname cnt@rev.desc.d2.uiv.p1.C.F.grid2-d3.n-apps\endcsname{12}
\expandafter\def\csname pct@rev.desc.d2.uiv.p1.C.F.grid2-d3.n-apps\endcsname{1200.0}
\expandafter\def\csname ppt@rev.desc.d2.uiv.p1.C.F.grid2-d3.n-apps\endcsname{$+$1200.0}
\expandafter\def\csname val@rev.desc.d2.uiv.p1.C.F.grid2-d3.n01\endcsname{25}
\expandafter\def\csname cnt@rev.desc.d2.uiv.p1.C.F.grid2-d3.n01\endcsname{25}
\expandafter\def\csname pct@rev.desc.d2.uiv.p1.C.F.grid2-d3.n01\endcsname{2500.0}
\expandafter\def\csname ppt@rev.desc.d2.uiv.p1.C.F.grid2-d3.n01\endcsname{$+$2500.0}
\expandafter\def\csname val@rev.desc.d2.uiv.p1.C.F.grid2-d3.n10\endcsname{2}
\expandafter\def\csname cnt@rev.desc.d2.uiv.p1.C.F.grid2-d3.n10\endcsname{2}
\expandafter\def\csname pct@rev.desc.d2.uiv.p1.C.F.grid2-d3.n10\endcsname{200.0}
\expandafter\def\csname ppt@rev.desc.d2.uiv.p1.C.F.grid2-d3.n10\endcsname{$+$200.0}
\expandafter\def\csname val@rev.desc.d2.uiv.p1.C.F.grid2-d3.rate-any\endcsname{0.281}
\expandafter\def\csname cnt@rev.desc.d2.uiv.p1.C.F.grid2-d3.rate-any\endcsname{0}
\expandafter\def\csname pct@rev.desc.d2.uiv.p1.C.F.grid2-d3.rate-any\endcsname{28.1}
\expandafter\def\csname ppt@rev.desc.d2.uiv.p1.C.F.grid2-d3.rate-any\endcsname{$+$28.1}
\expandafter\def\csname val@rev.desc.d2.uiv.p1.C.F.grid2-d3.rate01\endcsname{0.260}
\expandafter\def\csname cnt@rev.desc.d2.uiv.p1.C.F.grid2-d3.rate01\endcsname{0}
\expandafter\def\csname pct@rev.desc.d2.uiv.p1.C.F.grid2-d3.rate01\endcsname{26.0}
\expandafter\def\csname ppt@rev.desc.d2.uiv.p1.C.F.grid2-d3.rate01\endcsname{$+$26.0}
\expandafter\def\csname val@rev.desc.d2.uiv.p1.C.F.grid2-d3.rate10\endcsname{0.021}
\expandafter\def\csname cnt@rev.desc.d2.uiv.p1.C.F.grid2-d3.rate10\endcsname{0}
\expandafter\def\csname pct@rev.desc.d2.uiv.p1.C.F.grid2-d3.rate10\endcsname{2.1}
\expandafter\def\csname ppt@rev.desc.d2.uiv.p1.C.F.grid2-d3.rate10\endcsname{$+$2.1}
\expandafter\def\csname val@rev.desc.d2.uiv.p1.C.F.zoomclick-d3.acc-a\endcsname{0.604}
\expandafter\def\csname cnt@rev.desc.d2.uiv.p1.C.F.zoomclick-d3.acc-a\endcsname{1}
\expandafter\def\csname pct@rev.desc.d2.uiv.p1.C.F.zoomclick-d3.acc-a\endcsname{60.4}
\expandafter\def\csname ppt@rev.desc.d2.uiv.p1.C.F.zoomclick-d3.acc-a\endcsname{$+$60.4}
\expandafter\def\csname val@rev.desc.d2.uiv.p1.C.F.zoomclick-d3.acc-b\endcsname{0.740}
\expandafter\def\csname cnt@rev.desc.d2.uiv.p1.C.F.zoomclick-d3.acc-b\endcsname{1}
\expandafter\def\csname pct@rev.desc.d2.uiv.p1.C.F.zoomclick-d3.acc-b\endcsname{74.0}
\expandafter\def\csname ppt@rev.desc.d2.uiv.p1.C.F.zoomclick-d3.acc-b\endcsname{$+$74.0}
\expandafter\def\csname val@rev.desc.d2.uiv.p1.C.F.zoomclick-d3.cond01\endcsname{0.368}
\expandafter\def\csname cnt@rev.desc.d2.uiv.p1.C.F.zoomclick-d3.cond01\endcsname{0}
\expandafter\def\csname pct@rev.desc.d2.uiv.p1.C.F.zoomclick-d3.cond01\endcsname{36.8}
\expandafter\def\csname ppt@rev.desc.d2.uiv.p1.C.F.zoomclick-d3.cond01\endcsname{$+$36.8}
\expandafter\def\csname val@rev.desc.d2.uiv.p1.C.F.zoomclick-d3.cond10\endcsname{0.017}
\expandafter\def\csname cnt@rev.desc.d2.uiv.p1.C.F.zoomclick-d3.cond10\endcsname{0}
\expandafter\def\csname pct@rev.desc.d2.uiv.p1.C.F.zoomclick-d3.cond10\endcsname{1.7}
\expandafter\def\csname ppt@rev.desc.d2.uiv.p1.C.F.zoomclick-d3.cond10\endcsname{$+$1.7}
\expandafter\def\csname val@rev.desc.d2.uiv.p1.C.F.zoomclick-d3.n\endcsname{96}
\expandafter\def\csname cnt@rev.desc.d2.uiv.p1.C.F.zoomclick-d3.n\endcsname{96}
\expandafter\def\csname pct@rev.desc.d2.uiv.p1.C.F.zoomclick-d3.n\endcsname{9600.0}
\expandafter\def\csname ppt@rev.desc.d2.uiv.p1.C.F.zoomclick-d3.n\endcsname{$+$9600.0}
\expandafter\def\csname val@rev.desc.d2.uiv.p1.C.F.zoomclick-d3.n-apps\endcsname{12}
\expandafter\def\csname cnt@rev.desc.d2.uiv.p1.C.F.zoomclick-d3.n-apps\endcsname{12}
\expandafter\def\csname pct@rev.desc.d2.uiv.p1.C.F.zoomclick-d3.n-apps\endcsname{1200.0}
\expandafter\def\csname ppt@rev.desc.d2.uiv.p1.C.F.zoomclick-d3.n-apps\endcsname{$+$1200.0}
\expandafter\def\csname val@rev.desc.d2.uiv.p1.C.F.zoomclick-d3.n01\endcsname{14}
\expandafter\def\csname cnt@rev.desc.d2.uiv.p1.C.F.zoomclick-d3.n01\endcsname{14}
\expandafter\def\csname pct@rev.desc.d2.uiv.p1.C.F.zoomclick-d3.n01\endcsname{1400.0}
\expandafter\def\csname ppt@rev.desc.d2.uiv.p1.C.F.zoomclick-d3.n01\endcsname{$+$1400.0}
\expandafter\def\csname val@rev.desc.d2.uiv.p1.C.F.zoomclick-d3.n10\endcsname{1}
\expandafter\def\csname cnt@rev.desc.d2.uiv.p1.C.F.zoomclick-d3.n10\endcsname{1}
\expandafter\def\csname pct@rev.desc.d2.uiv.p1.C.F.zoomclick-d3.n10\endcsname{100.0}
\expandafter\def\csname ppt@rev.desc.d2.uiv.p1.C.F.zoomclick-d3.n10\endcsname{$+$100.0}
\expandafter\def\csname val@rev.desc.d2.uiv.p1.C.F.zoomclick-d3.rate-any\endcsname{0.156}
\expandafter\def\csname cnt@rev.desc.d2.uiv.p1.C.F.zoomclick-d3.rate-any\endcsname{0}
\expandafter\def\csname pct@rev.desc.d2.uiv.p1.C.F.zoomclick-d3.rate-any\endcsname{15.6}
\expandafter\def\csname ppt@rev.desc.d2.uiv.p1.C.F.zoomclick-d3.rate-any\endcsname{$+$15.6}
\expandafter\def\csname val@rev.desc.d2.uiv.p1.C.F.zoomclick-d3.rate01\endcsname{0.146}
\expandafter\def\csname cnt@rev.desc.d2.uiv.p1.C.F.zoomclick-d3.rate01\endcsname{0}
\expandafter\def\csname pct@rev.desc.d2.uiv.p1.C.F.zoomclick-d3.rate01\endcsname{14.6}
\expandafter\def\csname ppt@rev.desc.d2.uiv.p1.C.F.zoomclick-d3.rate01\endcsname{$+$14.6}
\expandafter\def\csname val@rev.desc.d2.uiv.p1.C.F.zoomclick-d3.rate10\endcsname{0.010}
\expandafter\def\csname cnt@rev.desc.d2.uiv.p1.C.F.zoomclick-d3.rate10\endcsname{0}
\expandafter\def\csname pct@rev.desc.d2.uiv.p1.C.F.zoomclick-d3.rate10\endcsname{1.0}
\expandafter\def\csname ppt@rev.desc.d2.uiv.p1.C.F.zoomclick-d3.rate10\endcsname{$+$1.0}
\expandafter\def\csname val@rev.desc.d2.uiv.p1.T.D.acc-a\endcsname{0.674}
\expandafter\def\csname cnt@rev.desc.d2.uiv.p1.T.D.acc-a\endcsname{1}
\expandafter\def\csname pct@rev.desc.d2.uiv.p1.T.D.acc-a\endcsname{67.4}
\expandafter\def\csname ppt@rev.desc.d2.uiv.p1.T.D.acc-a\endcsname{$+$67.4}
\expandafter\def\csname val@rev.desc.d2.uiv.p1.T.D.acc-b\endcsname{0.833}
\expandafter\def\csname cnt@rev.desc.d2.uiv.p1.T.D.acc-b\endcsname{1}
\expandafter\def\csname pct@rev.desc.d2.uiv.p1.T.D.acc-b\endcsname{83.3}
\expandafter\def\csname ppt@rev.desc.d2.uiv.p1.T.D.acc-b\endcsname{$+$83.3}
\expandafter\def\csname val@rev.desc.d2.uiv.p1.T.D.cond01\endcsname{0.553}
\expandafter\def\csname cnt@rev.desc.d2.uiv.p1.T.D.cond01\endcsname{1}
\expandafter\def\csname pct@rev.desc.d2.uiv.p1.T.D.cond01\endcsname{55.3}
\expandafter\def\csname ppt@rev.desc.d2.uiv.p1.T.D.cond01\endcsname{$+$55.3}
\expandafter\def\csname val@rev.desc.d2.uiv.p1.T.D.cond10\endcsname{0.031}
\expandafter\def\csname cnt@rev.desc.d2.uiv.p1.T.D.cond10\endcsname{0}
\expandafter\def\csname pct@rev.desc.d2.uiv.p1.T.D.cond10\endcsname{3.1}
\expandafter\def\csname ppt@rev.desc.d2.uiv.p1.T.D.cond10\endcsname{$+$3.1}
\expandafter\def\csname val@rev.desc.d2.uiv.p1.T.D.n\endcsname{144}
\expandafter\def\csname cnt@rev.desc.d2.uiv.p1.T.D.n\endcsname{144}
\expandafter\def\csname pct@rev.desc.d2.uiv.p1.T.D.n\endcsname{14400.0}
\expandafter\def\csname ppt@rev.desc.d2.uiv.p1.T.D.n\endcsname{$+$14400.0}
\expandafter\def\csname val@rev.desc.d2.uiv.p1.T.D.n-apps\endcsname{24}
\expandafter\def\csname cnt@rev.desc.d2.uiv.p1.T.D.n-apps\endcsname{24}
\expandafter\def\csname pct@rev.desc.d2.uiv.p1.T.D.n-apps\endcsname{2400.0}
\expandafter\def\csname ppt@rev.desc.d2.uiv.p1.T.D.n-apps\endcsname{$+$2400.0}
\expandafter\def\csname val@rev.desc.d2.uiv.p1.T.D.n01\endcsname{26}
\expandafter\def\csname cnt@rev.desc.d2.uiv.p1.T.D.n01\endcsname{26}
\expandafter\def\csname pct@rev.desc.d2.uiv.p1.T.D.n01\endcsname{2600.0}
\expandafter\def\csname ppt@rev.desc.d2.uiv.p1.T.D.n01\endcsname{$+$2600.0}
\expandafter\def\csname val@rev.desc.d2.uiv.p1.T.D.n10\endcsname{3}
\expandafter\def\csname cnt@rev.desc.d2.uiv.p1.T.D.n10\endcsname{3}
\expandafter\def\csname pct@rev.desc.d2.uiv.p1.T.D.n10\endcsname{300.0}
\expandafter\def\csname ppt@rev.desc.d2.uiv.p1.T.D.n10\endcsname{$+$300.0}
\expandafter\def\csname val@rev.desc.d2.uiv.p1.T.D.rate-any\endcsname{0.201}
\expandafter\def\csname cnt@rev.desc.d2.uiv.p1.T.D.rate-any\endcsname{0}
\expandafter\def\csname pct@rev.desc.d2.uiv.p1.T.D.rate-any\endcsname{20.1}
\expandafter\def\csname ppt@rev.desc.d2.uiv.p1.T.D.rate-any\endcsname{$+$20.1}
\expandafter\def\csname val@rev.desc.d2.uiv.p1.T.D.rate01\endcsname{0.181}
\expandafter\def\csname cnt@rev.desc.d2.uiv.p1.T.D.rate01\endcsname{0}
\expandafter\def\csname pct@rev.desc.d2.uiv.p1.T.D.rate01\endcsname{18.1}
\expandafter\def\csname ppt@rev.desc.d2.uiv.p1.T.D.rate01\endcsname{$+$18.1}
\expandafter\def\csname val@rev.desc.d2.uiv.p1.T.D.rate10\endcsname{0.021}
\expandafter\def\csname cnt@rev.desc.d2.uiv.p1.T.D.rate10\endcsname{0}
\expandafter\def\csname pct@rev.desc.d2.uiv.p1.T.D.rate10\endcsname{2.1}
\expandafter\def\csname ppt@rev.desc.d2.uiv.p1.T.D.rate10\endcsname{$+$2.1}
\expandafter\def\csname val@rev.desc.d2.uiv.p2.C.D.acc-a\endcsname{0.760}
\expandafter\def\csname cnt@rev.desc.d2.uiv.p2.C.D.acc-a\endcsname{1}
\expandafter\def\csname pct@rev.desc.d2.uiv.p2.C.D.acc-a\endcsname{76.0}
\expandafter\def\csname ppt@rev.desc.d2.uiv.p2.C.D.acc-a\endcsname{$+$76.0}
\expandafter\def\csname val@rev.desc.d2.uiv.p2.C.D.acc-b\endcsname{0.792}
\expandafter\def\csname cnt@rev.desc.d2.uiv.p2.C.D.acc-b\endcsname{1}
\expandafter\def\csname pct@rev.desc.d2.uiv.p2.C.D.acc-b\endcsname{79.2}
\expandafter\def\csname ppt@rev.desc.d2.uiv.p2.C.D.acc-b\endcsname{$+$79.2}
\expandafter\def\csname val@rev.desc.d2.uiv.p2.C.D.cond01\endcsname{0.304}
\expandafter\def\csname cnt@rev.desc.d2.uiv.p2.C.D.cond01\endcsname{0}
\expandafter\def\csname pct@rev.desc.d2.uiv.p2.C.D.cond01\endcsname{30.4}
\expandafter\def\csname ppt@rev.desc.d2.uiv.p2.C.D.cond01\endcsname{$+$30.4}
\expandafter\def\csname val@rev.desc.d2.uiv.p2.C.D.cond10\endcsname{0.055}
\expandafter\def\csname cnt@rev.desc.d2.uiv.p2.C.D.cond10\endcsname{0}
\expandafter\def\csname pct@rev.desc.d2.uiv.p2.C.D.cond10\endcsname{5.5}
\expandafter\def\csname ppt@rev.desc.d2.uiv.p2.C.D.cond10\endcsname{$+$5.5}
\expandafter\def\csname val@rev.desc.d2.uiv.p2.C.D.n\endcsname{96}
\expandafter\def\csname cnt@rev.desc.d2.uiv.p2.C.D.n\endcsname{96}
\expandafter\def\csname pct@rev.desc.d2.uiv.p2.C.D.n\endcsname{9600.0}
\expandafter\def\csname ppt@rev.desc.d2.uiv.p2.C.D.n\endcsname{$+$9600.0}
\expandafter\def\csname val@rev.desc.d2.uiv.p2.C.D.n-apps\endcsname{12}
\expandafter\def\csname cnt@rev.desc.d2.uiv.p2.C.D.n-apps\endcsname{12}
\expandafter\def\csname pct@rev.desc.d2.uiv.p2.C.D.n-apps\endcsname{1200.0}
\expandafter\def\csname ppt@rev.desc.d2.uiv.p2.C.D.n-apps\endcsname{$+$1200.0}
\expandafter\def\csname val@rev.desc.d2.uiv.p2.C.D.n01\endcsname{7}
\expandafter\def\csname cnt@rev.desc.d2.uiv.p2.C.D.n01\endcsname{7}
\expandafter\def\csname pct@rev.desc.d2.uiv.p2.C.D.n01\endcsname{700.0}
\expandafter\def\csname ppt@rev.desc.d2.uiv.p2.C.D.n01\endcsname{$+$700.0}
\expandafter\def\csname val@rev.desc.d2.uiv.p2.C.D.n10\endcsname{4}
\expandafter\def\csname cnt@rev.desc.d2.uiv.p2.C.D.n10\endcsname{4}
\expandafter\def\csname pct@rev.desc.d2.uiv.p2.C.D.n10\endcsname{400.0}
\expandafter\def\csname ppt@rev.desc.d2.uiv.p2.C.D.n10\endcsname{$+$400.0}
\expandafter\def\csname val@rev.desc.d2.uiv.p2.C.D.rate-any\endcsname{0.115}
\expandafter\def\csname cnt@rev.desc.d2.uiv.p2.C.D.rate-any\endcsname{0}
\expandafter\def\csname pct@rev.desc.d2.uiv.p2.C.D.rate-any\endcsname{11.5}
\expandafter\def\csname ppt@rev.desc.d2.uiv.p2.C.D.rate-any\endcsname{$+$11.5}
\expandafter\def\csname val@rev.desc.d2.uiv.p2.C.D.rate01\endcsname{0.073}
\expandafter\def\csname cnt@rev.desc.d2.uiv.p2.C.D.rate01\endcsname{0}
\expandafter\def\csname pct@rev.desc.d2.uiv.p2.C.D.rate01\endcsname{7.3}
\expandafter\def\csname ppt@rev.desc.d2.uiv.p2.C.D.rate01\endcsname{$+$7.3}
\expandafter\def\csname val@rev.desc.d2.uiv.p2.C.D.rate10\endcsname{0.042}
\expandafter\def\csname cnt@rev.desc.d2.uiv.p2.C.D.rate10\endcsname{0}
\expandafter\def\csname pct@rev.desc.d2.uiv.p2.C.D.rate10\endcsname{4.2}
\expandafter\def\csname ppt@rev.desc.d2.uiv.p2.C.D.rate10\endcsname{$+$4.2}
\expandafter\def\csname val@rev.desc.d2.uiv.p2.C.F.center-d1.acc-a\endcsname{0.771}
\expandafter\def\csname cnt@rev.desc.d2.uiv.p2.C.F.center-d1.acc-a\endcsname{1}
\expandafter\def\csname pct@rev.desc.d2.uiv.p2.C.F.center-d1.acc-a\endcsname{77.1}
\expandafter\def\csname ppt@rev.desc.d2.uiv.p2.C.F.center-d1.acc-a\endcsname{$+$77.1}
\expandafter\def\csname val@rev.desc.d2.uiv.p2.C.F.center-d1.acc-b\endcsname{0.812}
\expandafter\def\csname cnt@rev.desc.d2.uiv.p2.C.F.center-d1.acc-b\endcsname{1}
\expandafter\def\csname pct@rev.desc.d2.uiv.p2.C.F.center-d1.acc-b\endcsname{81.2}
\expandafter\def\csname ppt@rev.desc.d2.uiv.p2.C.F.center-d1.acc-b\endcsname{$+$81.2}
\expandafter\def\csname val@rev.desc.d2.uiv.p2.C.F.center-d1.cond01\endcsname{0.273}
\expandafter\def\csname cnt@rev.desc.d2.uiv.p2.C.F.center-d1.cond01\endcsname{0}
\expandafter\def\csname pct@rev.desc.d2.uiv.p2.C.F.center-d1.cond01\endcsname{27.3}
\expandafter\def\csname ppt@rev.desc.d2.uiv.p2.C.F.center-d1.cond01\endcsname{$+$27.3}
\expandafter\def\csname val@rev.desc.d2.uiv.p2.C.F.center-d1.cond10\endcsname{0.027}
\expandafter\def\csname cnt@rev.desc.d2.uiv.p2.C.F.center-d1.cond10\endcsname{0}
\expandafter\def\csname pct@rev.desc.d2.uiv.p2.C.F.center-d1.cond10\endcsname{2.7}
\expandafter\def\csname ppt@rev.desc.d2.uiv.p2.C.F.center-d1.cond10\endcsname{$+$2.7}
\expandafter\def\csname val@rev.desc.d2.uiv.p2.C.F.center-d1.n\endcsname{96}
\expandafter\def\csname cnt@rev.desc.d2.uiv.p2.C.F.center-d1.n\endcsname{96}
\expandafter\def\csname pct@rev.desc.d2.uiv.p2.C.F.center-d1.n\endcsname{9600.0}
\expandafter\def\csname ppt@rev.desc.d2.uiv.p2.C.F.center-d1.n\endcsname{$+$9600.0}
\expandafter\def\csname val@rev.desc.d2.uiv.p2.C.F.center-d1.n-apps\endcsname{12}
\expandafter\def\csname cnt@rev.desc.d2.uiv.p2.C.F.center-d1.n-apps\endcsname{12}
\expandafter\def\csname pct@rev.desc.d2.uiv.p2.C.F.center-d1.n-apps\endcsname{1200.0}
\expandafter\def\csname ppt@rev.desc.d2.uiv.p2.C.F.center-d1.n-apps\endcsname{$+$1200.0}
\expandafter\def\csname val@rev.desc.d2.uiv.p2.C.F.center-d1.n01\endcsname{6}
\expandafter\def\csname cnt@rev.desc.d2.uiv.p2.C.F.center-d1.n01\endcsname{6}
\expandafter\def\csname pct@rev.desc.d2.uiv.p2.C.F.center-d1.n01\endcsname{600.0}
\expandafter\def\csname ppt@rev.desc.d2.uiv.p2.C.F.center-d1.n01\endcsname{$+$600.0}
\expandafter\def\csname val@rev.desc.d2.uiv.p2.C.F.center-d1.n10\endcsname{2}
\expandafter\def\csname cnt@rev.desc.d2.uiv.p2.C.F.center-d1.n10\endcsname{2}
\expandafter\def\csname pct@rev.desc.d2.uiv.p2.C.F.center-d1.n10\endcsname{200.0}
\expandafter\def\csname ppt@rev.desc.d2.uiv.p2.C.F.center-d1.n10\endcsname{$+$200.0}
\expandafter\def\csname val@rev.desc.d2.uiv.p2.C.F.center-d1.rate-any\endcsname{0.083}
\expandafter\def\csname cnt@rev.desc.d2.uiv.p2.C.F.center-d1.rate-any\endcsname{0}
\expandafter\def\csname pct@rev.desc.d2.uiv.p2.C.F.center-d1.rate-any\endcsname{8.3}
\expandafter\def\csname ppt@rev.desc.d2.uiv.p2.C.F.center-d1.rate-any\endcsname{$+$8.3}
\expandafter\def\csname val@rev.desc.d2.uiv.p2.C.F.center-d1.rate01\endcsname{0.062}
\expandafter\def\csname cnt@rev.desc.d2.uiv.p2.C.F.center-d1.rate01\endcsname{0}
\expandafter\def\csname pct@rev.desc.d2.uiv.p2.C.F.center-d1.rate01\endcsname{6.2}
\expandafter\def\csname ppt@rev.desc.d2.uiv.p2.C.F.center-d1.rate01\endcsname{$+$6.2}
\expandafter\def\csname val@rev.desc.d2.uiv.p2.C.F.center-d1.rate10\endcsname{0.021}
\expandafter\def\csname cnt@rev.desc.d2.uiv.p2.C.F.center-d1.rate10\endcsname{0}
\expandafter\def\csname pct@rev.desc.d2.uiv.p2.C.F.center-d1.rate10\endcsname{2.1}
\expandafter\def\csname ppt@rev.desc.d2.uiv.p2.C.F.center-d1.rate10\endcsname{$+$2.1}
\expandafter\def\csname val@rev.desc.d2.uiv.p2.C.F.context-d1.acc-a\endcsname{0.740}
\expandafter\def\csname cnt@rev.desc.d2.uiv.p2.C.F.context-d1.acc-a\endcsname{1}
\expandafter\def\csname pct@rev.desc.d2.uiv.p2.C.F.context-d1.acc-a\endcsname{74.0}
\expandafter\def\csname ppt@rev.desc.d2.uiv.p2.C.F.context-d1.acc-a\endcsname{$+$74.0}
\expandafter\def\csname val@rev.desc.d2.uiv.p2.C.F.context-d1.acc-b\endcsname{0.802}
\expandafter\def\csname cnt@rev.desc.d2.uiv.p2.C.F.context-d1.acc-b\endcsname{1}
\expandafter\def\csname pct@rev.desc.d2.uiv.p2.C.F.context-d1.acc-b\endcsname{80.2}
\expandafter\def\csname ppt@rev.desc.d2.uiv.p2.C.F.context-d1.acc-b\endcsname{$+$80.2}
\expandafter\def\csname val@rev.desc.d2.uiv.p2.C.F.context-d1.cond01\endcsname{0.360}
\expandafter\def\csname cnt@rev.desc.d2.uiv.p2.C.F.context-d1.cond01\endcsname{0}
\expandafter\def\csname pct@rev.desc.d2.uiv.p2.C.F.context-d1.cond01\endcsname{36.0}
\expandafter\def\csname ppt@rev.desc.d2.uiv.p2.C.F.context-d1.cond01\endcsname{$+$36.0}
\expandafter\def\csname val@rev.desc.d2.uiv.p2.C.F.context-d1.cond10\endcsname{0.042}
\expandafter\def\csname cnt@rev.desc.d2.uiv.p2.C.F.context-d1.cond10\endcsname{0}
\expandafter\def\csname pct@rev.desc.d2.uiv.p2.C.F.context-d1.cond10\endcsname{4.2}
\expandafter\def\csname ppt@rev.desc.d2.uiv.p2.C.F.context-d1.cond10\endcsname{$+$4.2}
\expandafter\def\csname val@rev.desc.d2.uiv.p2.C.F.context-d1.n\endcsname{96}
\expandafter\def\csname cnt@rev.desc.d2.uiv.p2.C.F.context-d1.n\endcsname{96}
\expandafter\def\csname pct@rev.desc.d2.uiv.p2.C.F.context-d1.n\endcsname{9600.0}
\expandafter\def\csname ppt@rev.desc.d2.uiv.p2.C.F.context-d1.n\endcsname{$+$9600.0}
\expandafter\def\csname val@rev.desc.d2.uiv.p2.C.F.context-d1.n-apps\endcsname{12}
\expandafter\def\csname cnt@rev.desc.d2.uiv.p2.C.F.context-d1.n-apps\endcsname{12}
\expandafter\def\csname pct@rev.desc.d2.uiv.p2.C.F.context-d1.n-apps\endcsname{1200.0}
\expandafter\def\csname ppt@rev.desc.d2.uiv.p2.C.F.context-d1.n-apps\endcsname{$+$1200.0}
\expandafter\def\csname val@rev.desc.d2.uiv.p2.C.F.context-d1.n01\endcsname{9}
\expandafter\def\csname cnt@rev.desc.d2.uiv.p2.C.F.context-d1.n01\endcsname{9}
\expandafter\def\csname pct@rev.desc.d2.uiv.p2.C.F.context-d1.n01\endcsname{900.0}
\expandafter\def\csname ppt@rev.desc.d2.uiv.p2.C.F.context-d1.n01\endcsname{$+$900.0}
\expandafter\def\csname val@rev.desc.d2.uiv.p2.C.F.context-d1.n10\endcsname{3}
\expandafter\def\csname cnt@rev.desc.d2.uiv.p2.C.F.context-d1.n10\endcsname{3}
\expandafter\def\csname pct@rev.desc.d2.uiv.p2.C.F.context-d1.n10\endcsname{300.0}
\expandafter\def\csname ppt@rev.desc.d2.uiv.p2.C.F.context-d1.n10\endcsname{$+$300.0}
\expandafter\def\csname val@rev.desc.d2.uiv.p2.C.F.context-d1.rate-any\endcsname{0.125}
\expandafter\def\csname cnt@rev.desc.d2.uiv.p2.C.F.context-d1.rate-any\endcsname{0}
\expandafter\def\csname pct@rev.desc.d2.uiv.p2.C.F.context-d1.rate-any\endcsname{12.5}
\expandafter\def\csname ppt@rev.desc.d2.uiv.p2.C.F.context-d1.rate-any\endcsname{$+$12.5}
\expandafter\def\csname val@rev.desc.d2.uiv.p2.C.F.context-d1.rate01\endcsname{0.094}
\expandafter\def\csname cnt@rev.desc.d2.uiv.p2.C.F.context-d1.rate01\endcsname{0}
\expandafter\def\csname pct@rev.desc.d2.uiv.p2.C.F.context-d1.rate01\endcsname{9.4}
\expandafter\def\csname ppt@rev.desc.d2.uiv.p2.C.F.context-d1.rate01\endcsname{$+$9.4}
\expandafter\def\csname val@rev.desc.d2.uiv.p2.C.F.context-d1.rate10\endcsname{0.031}
\expandafter\def\csname cnt@rev.desc.d2.uiv.p2.C.F.context-d1.rate10\endcsname{0}
\expandafter\def\csname pct@rev.desc.d2.uiv.p2.C.F.context-d1.rate10\endcsname{3.1}
\expandafter\def\csname ppt@rev.desc.d2.uiv.p2.C.F.context-d1.rate10\endcsname{$+$3.1}
\expandafter\def\csname val@rev.desc.d2.uiv.p2.C.F.dual-d2.acc-a\endcsname{0.760}
\expandafter\def\csname cnt@rev.desc.d2.uiv.p2.C.F.dual-d2.acc-a\endcsname{1}
\expandafter\def\csname pct@rev.desc.d2.uiv.p2.C.F.dual-d2.acc-a\endcsname{76.0}
\expandafter\def\csname ppt@rev.desc.d2.uiv.p2.C.F.dual-d2.acc-a\endcsname{$+$76.0}
\expandafter\def\csname val@rev.desc.d2.uiv.p2.C.F.dual-d2.acc-b\endcsname{0.781}
\expandafter\def\csname cnt@rev.desc.d2.uiv.p2.C.F.dual-d2.acc-b\endcsname{1}
\expandafter\def\csname pct@rev.desc.d2.uiv.p2.C.F.dual-d2.acc-b\endcsname{78.1}
\expandafter\def\csname ppt@rev.desc.d2.uiv.p2.C.F.dual-d2.acc-b\endcsname{$+$78.1}
\expandafter\def\csname val@rev.desc.d2.uiv.p2.C.F.dual-d2.cond01\endcsname{0.217}
\expandafter\def\csname cnt@rev.desc.d2.uiv.p2.C.F.dual-d2.cond01\endcsname{0}
\expandafter\def\csname pct@rev.desc.d2.uiv.p2.C.F.dual-d2.cond01\endcsname{21.7}
\expandafter\def\csname ppt@rev.desc.d2.uiv.p2.C.F.dual-d2.cond01\endcsname{$+$21.7}
\expandafter\def\csname val@rev.desc.d2.uiv.p2.C.F.dual-d2.cond10\endcsname{0.041}
\expandafter\def\csname cnt@rev.desc.d2.uiv.p2.C.F.dual-d2.cond10\endcsname{0}
\expandafter\def\csname pct@rev.desc.d2.uiv.p2.C.F.dual-d2.cond10\endcsname{4.1}
\expandafter\def\csname ppt@rev.desc.d2.uiv.p2.C.F.dual-d2.cond10\endcsname{$+$4.1}
\expandafter\def\csname val@rev.desc.d2.uiv.p2.C.F.dual-d2.n\endcsname{96}
\expandafter\def\csname cnt@rev.desc.d2.uiv.p2.C.F.dual-d2.n\endcsname{96}
\expandafter\def\csname pct@rev.desc.d2.uiv.p2.C.F.dual-d2.n\endcsname{9600.0}
\expandafter\def\csname ppt@rev.desc.d2.uiv.p2.C.F.dual-d2.n\endcsname{$+$9600.0}
\expandafter\def\csname val@rev.desc.d2.uiv.p2.C.F.dual-d2.n-apps\endcsname{12}
\expandafter\def\csname cnt@rev.desc.d2.uiv.p2.C.F.dual-d2.n-apps\endcsname{12}
\expandafter\def\csname pct@rev.desc.d2.uiv.p2.C.F.dual-d2.n-apps\endcsname{1200.0}
\expandafter\def\csname ppt@rev.desc.d2.uiv.p2.C.F.dual-d2.n-apps\endcsname{$+$1200.0}
\expandafter\def\csname val@rev.desc.d2.uiv.p2.C.F.dual-d2.n01\endcsname{5}
\expandafter\def\csname cnt@rev.desc.d2.uiv.p2.C.F.dual-d2.n01\endcsname{5}
\expandafter\def\csname pct@rev.desc.d2.uiv.p2.C.F.dual-d2.n01\endcsname{500.0}
\expandafter\def\csname ppt@rev.desc.d2.uiv.p2.C.F.dual-d2.n01\endcsname{$+$500.0}
\expandafter\def\csname val@rev.desc.d2.uiv.p2.C.F.dual-d2.n10\endcsname{3}
\expandafter\def\csname cnt@rev.desc.d2.uiv.p2.C.F.dual-d2.n10\endcsname{3}
\expandafter\def\csname pct@rev.desc.d2.uiv.p2.C.F.dual-d2.n10\endcsname{300.0}
\expandafter\def\csname ppt@rev.desc.d2.uiv.p2.C.F.dual-d2.n10\endcsname{$+$300.0}
\expandafter\def\csname val@rev.desc.d2.uiv.p2.C.F.dual-d2.rate-any\endcsname{0.083}
\expandafter\def\csname cnt@rev.desc.d2.uiv.p2.C.F.dual-d2.rate-any\endcsname{0}
\expandafter\def\csname pct@rev.desc.d2.uiv.p2.C.F.dual-d2.rate-any\endcsname{8.3}
\expandafter\def\csname ppt@rev.desc.d2.uiv.p2.C.F.dual-d2.rate-any\endcsname{$+$8.3}
\expandafter\def\csname val@rev.desc.d2.uiv.p2.C.F.dual-d2.rate01\endcsname{0.052}
\expandafter\def\csname cnt@rev.desc.d2.uiv.p2.C.F.dual-d2.rate01\endcsname{0}
\expandafter\def\csname pct@rev.desc.d2.uiv.p2.C.F.dual-d2.rate01\endcsname{5.2}
\expandafter\def\csname ppt@rev.desc.d2.uiv.p2.C.F.dual-d2.rate01\endcsname{$+$5.2}
\expandafter\def\csname val@rev.desc.d2.uiv.p2.C.F.dual-d2.rate10\endcsname{0.031}
\expandafter\def\csname cnt@rev.desc.d2.uiv.p2.C.F.dual-d2.rate10\endcsname{0}
\expandafter\def\csname pct@rev.desc.d2.uiv.p2.C.F.dual-d2.rate10\endcsname{3.1}
\expandafter\def\csname ppt@rev.desc.d2.uiv.p2.C.F.dual-d2.rate10\endcsname{$+$3.1}
\expandafter\def\csname val@rev.desc.d2.uiv.p2.C.F.grid2-d3.acc-a\endcsname{0.729}
\expandafter\def\csname cnt@rev.desc.d2.uiv.p2.C.F.grid2-d3.acc-a\endcsname{1}
\expandafter\def\csname pct@rev.desc.d2.uiv.p2.C.F.grid2-d3.acc-a\endcsname{72.9}
\expandafter\def\csname ppt@rev.desc.d2.uiv.p2.C.F.grid2-d3.acc-a\endcsname{$+$72.9}
\expandafter\def\csname val@rev.desc.d2.uiv.p2.C.F.grid2-d3.acc-b\endcsname{0.802}
\expandafter\def\csname cnt@rev.desc.d2.uiv.p2.C.F.grid2-d3.acc-b\endcsname{1}
\expandafter\def\csname pct@rev.desc.d2.uiv.p2.C.F.grid2-d3.acc-b\endcsname{80.2}
\expandafter\def\csname ppt@rev.desc.d2.uiv.p2.C.F.grid2-d3.acc-b\endcsname{$+$80.2}
\expandafter\def\csname val@rev.desc.d2.uiv.p2.C.F.grid2-d3.cond01\endcsname{0.423}
\expandafter\def\csname cnt@rev.desc.d2.uiv.p2.C.F.grid2-d3.cond01\endcsname{0}
\expandafter\def\csname pct@rev.desc.d2.uiv.p2.C.F.grid2-d3.cond01\endcsname{42.3}
\expandafter\def\csname ppt@rev.desc.d2.uiv.p2.C.F.grid2-d3.cond01\endcsname{$+$42.3}
\expandafter\def\csname val@rev.desc.d2.uiv.p2.C.F.grid2-d3.cond10\endcsname{0.057}
\expandafter\def\csname cnt@rev.desc.d2.uiv.p2.C.F.grid2-d3.cond10\endcsname{0}
\expandafter\def\csname pct@rev.desc.d2.uiv.p2.C.F.grid2-d3.cond10\endcsname{5.7}
\expandafter\def\csname ppt@rev.desc.d2.uiv.p2.C.F.grid2-d3.cond10\endcsname{$+$5.7}
\expandafter\def\csname val@rev.desc.d2.uiv.p2.C.F.grid2-d3.n\endcsname{96}
\expandafter\def\csname cnt@rev.desc.d2.uiv.p2.C.F.grid2-d3.n\endcsname{96}
\expandafter\def\csname pct@rev.desc.d2.uiv.p2.C.F.grid2-d3.n\endcsname{9600.0}
\expandafter\def\csname ppt@rev.desc.d2.uiv.p2.C.F.grid2-d3.n\endcsname{$+$9600.0}
\expandafter\def\csname val@rev.desc.d2.uiv.p2.C.F.grid2-d3.n-apps\endcsname{12}
\expandafter\def\csname cnt@rev.desc.d2.uiv.p2.C.F.grid2-d3.n-apps\endcsname{12}
\expandafter\def\csname pct@rev.desc.d2.uiv.p2.C.F.grid2-d3.n-apps\endcsname{1200.0}
\expandafter\def\csname ppt@rev.desc.d2.uiv.p2.C.F.grid2-d3.n-apps\endcsname{$+$1200.0}
\expandafter\def\csname val@rev.desc.d2.uiv.p2.C.F.grid2-d3.n01\endcsname{11}
\expandafter\def\csname cnt@rev.desc.d2.uiv.p2.C.F.grid2-d3.n01\endcsname{11}
\expandafter\def\csname pct@rev.desc.d2.uiv.p2.C.F.grid2-d3.n01\endcsname{1100.0}
\expandafter\def\csname ppt@rev.desc.d2.uiv.p2.C.F.grid2-d3.n01\endcsname{$+$1100.0}
\expandafter\def\csname val@rev.desc.d2.uiv.p2.C.F.grid2-d3.n10\endcsname{4}
\expandafter\def\csname cnt@rev.desc.d2.uiv.p2.C.F.grid2-d3.n10\endcsname{4}
\expandafter\def\csname pct@rev.desc.d2.uiv.p2.C.F.grid2-d3.n10\endcsname{400.0}
\expandafter\def\csname ppt@rev.desc.d2.uiv.p2.C.F.grid2-d3.n10\endcsname{$+$400.0}
\expandafter\def\csname val@rev.desc.d2.uiv.p2.C.F.grid2-d3.rate-any\endcsname{0.156}
\expandafter\def\csname cnt@rev.desc.d2.uiv.p2.C.F.grid2-d3.rate-any\endcsname{0}
\expandafter\def\csname pct@rev.desc.d2.uiv.p2.C.F.grid2-d3.rate-any\endcsname{15.6}
\expandafter\def\csname ppt@rev.desc.d2.uiv.p2.C.F.grid2-d3.rate-any\endcsname{$+$15.6}
\expandafter\def\csname val@rev.desc.d2.uiv.p2.C.F.grid2-d3.rate01\endcsname{0.115}
\expandafter\def\csname cnt@rev.desc.d2.uiv.p2.C.F.grid2-d3.rate01\endcsname{0}
\expandafter\def\csname pct@rev.desc.d2.uiv.p2.C.F.grid2-d3.rate01\endcsname{11.5}
\expandafter\def\csname ppt@rev.desc.d2.uiv.p2.C.F.grid2-d3.rate01\endcsname{$+$11.5}
\expandafter\def\csname val@rev.desc.d2.uiv.p2.C.F.grid2-d3.rate10\endcsname{0.042}
\expandafter\def\csname cnt@rev.desc.d2.uiv.p2.C.F.grid2-d3.rate10\endcsname{0}
\expandafter\def\csname pct@rev.desc.d2.uiv.p2.C.F.grid2-d3.rate10\endcsname{4.2}
\expandafter\def\csname ppt@rev.desc.d2.uiv.p2.C.F.grid2-d3.rate10\endcsname{$+$4.2}
\expandafter\def\csname val@rev.desc.d2.uiv.p2.C.F.zoomclick-d3.acc-a\endcsname{0.740}
\expandafter\def\csname cnt@rev.desc.d2.uiv.p2.C.F.zoomclick-d3.acc-a\endcsname{1}
\expandafter\def\csname pct@rev.desc.d2.uiv.p2.C.F.zoomclick-d3.acc-a\endcsname{74.0}
\expandafter\def\csname ppt@rev.desc.d2.uiv.p2.C.F.zoomclick-d3.acc-a\endcsname{$+$74.0}
\expandafter\def\csname val@rev.desc.d2.uiv.p2.C.F.zoomclick-d3.acc-b\endcsname{0.802}
\expandafter\def\csname cnt@rev.desc.d2.uiv.p2.C.F.zoomclick-d3.acc-b\endcsname{1}
\expandafter\def\csname pct@rev.desc.d2.uiv.p2.C.F.zoomclick-d3.acc-b\endcsname{80.2}
\expandafter\def\csname ppt@rev.desc.d2.uiv.p2.C.F.zoomclick-d3.acc-b\endcsname{$+$80.2}
\expandafter\def\csname val@rev.desc.d2.uiv.p2.C.F.zoomclick-d3.cond01\endcsname{0.400}
\expandafter\def\csname cnt@rev.desc.d2.uiv.p2.C.F.zoomclick-d3.cond01\endcsname{0}
\expandafter\def\csname pct@rev.desc.d2.uiv.p2.C.F.zoomclick-d3.cond01\endcsname{40.0}
\expandafter\def\csname ppt@rev.desc.d2.uiv.p2.C.F.zoomclick-d3.cond01\endcsname{$+$40.0}
\expandafter\def\csname val@rev.desc.d2.uiv.p2.C.F.zoomclick-d3.cond10\endcsname{0.056}
\expandafter\def\csname cnt@rev.desc.d2.uiv.p2.C.F.zoomclick-d3.cond10\endcsname{0}
\expandafter\def\csname pct@rev.desc.d2.uiv.p2.C.F.zoomclick-d3.cond10\endcsname{5.6}
\expandafter\def\csname ppt@rev.desc.d2.uiv.p2.C.F.zoomclick-d3.cond10\endcsname{$+$5.6}
\expandafter\def\csname val@rev.desc.d2.uiv.p2.C.F.zoomclick-d3.n\endcsname{96}
\expandafter\def\csname cnt@rev.desc.d2.uiv.p2.C.F.zoomclick-d3.n\endcsname{96}
\expandafter\def\csname pct@rev.desc.d2.uiv.p2.C.F.zoomclick-d3.n\endcsname{9600.0}
\expandafter\def\csname ppt@rev.desc.d2.uiv.p2.C.F.zoomclick-d3.n\endcsname{$+$9600.0}
\expandafter\def\csname val@rev.desc.d2.uiv.p2.C.F.zoomclick-d3.n-apps\endcsname{12}
\expandafter\def\csname cnt@rev.desc.d2.uiv.p2.C.F.zoomclick-d3.n-apps\endcsname{12}
\expandafter\def\csname pct@rev.desc.d2.uiv.p2.C.F.zoomclick-d3.n-apps\endcsname{1200.0}
\expandafter\def\csname ppt@rev.desc.d2.uiv.p2.C.F.zoomclick-d3.n-apps\endcsname{$+$1200.0}
\expandafter\def\csname val@rev.desc.d2.uiv.p2.C.F.zoomclick-d3.n01\endcsname{10}
\expandafter\def\csname cnt@rev.desc.d2.uiv.p2.C.F.zoomclick-d3.n01\endcsname{10}
\expandafter\def\csname pct@rev.desc.d2.uiv.p2.C.F.zoomclick-d3.n01\endcsname{1000.0}
\expandafter\def\csname ppt@rev.desc.d2.uiv.p2.C.F.zoomclick-d3.n01\endcsname{$+$1000.0}
\expandafter\def\csname val@rev.desc.d2.uiv.p2.C.F.zoomclick-d3.n10\endcsname{4}
\expandafter\def\csname cnt@rev.desc.d2.uiv.p2.C.F.zoomclick-d3.n10\endcsname{4}
\expandafter\def\csname pct@rev.desc.d2.uiv.p2.C.F.zoomclick-d3.n10\endcsname{400.0}
\expandafter\def\csname ppt@rev.desc.d2.uiv.p2.C.F.zoomclick-d3.n10\endcsname{$+$400.0}
\expandafter\def\csname val@rev.desc.d2.uiv.p2.C.F.zoomclick-d3.rate-any\endcsname{0.146}
\expandafter\def\csname cnt@rev.desc.d2.uiv.p2.C.F.zoomclick-d3.rate-any\endcsname{0}
\expandafter\def\csname pct@rev.desc.d2.uiv.p2.C.F.zoomclick-d3.rate-any\endcsname{14.6}
\expandafter\def\csname ppt@rev.desc.d2.uiv.p2.C.F.zoomclick-d3.rate-any\endcsname{$+$14.6}
\expandafter\def\csname val@rev.desc.d2.uiv.p2.C.F.zoomclick-d3.rate01\endcsname{0.104}
\expandafter\def\csname cnt@rev.desc.d2.uiv.p2.C.F.zoomclick-d3.rate01\endcsname{0}
\expandafter\def\csname pct@rev.desc.d2.uiv.p2.C.F.zoomclick-d3.rate01\endcsname{10.4}
\expandafter\def\csname ppt@rev.desc.d2.uiv.p2.C.F.zoomclick-d3.rate01\endcsname{$+$10.4}
\expandafter\def\csname val@rev.desc.d2.uiv.p2.C.F.zoomclick-d3.rate10\endcsname{0.042}
\expandafter\def\csname cnt@rev.desc.d2.uiv.p2.C.F.zoomclick-d3.rate10\endcsname{0}
\expandafter\def\csname pct@rev.desc.d2.uiv.p2.C.F.zoomclick-d3.rate10\endcsname{4.2}
\expandafter\def\csname ppt@rev.desc.d2.uiv.p2.C.F.zoomclick-d3.rate10\endcsname{$+$4.2}
\expandafter\def\csname val@rev.desc.d2.uiv.p2.T.D.acc-a\endcsname{0.833}
\expandafter\def\csname cnt@rev.desc.d2.uiv.p2.T.D.acc-a\endcsname{1}
\expandafter\def\csname pct@rev.desc.d2.uiv.p2.T.D.acc-a\endcsname{83.3}
\expandafter\def\csname ppt@rev.desc.d2.uiv.p2.T.D.acc-a\endcsname{$+$83.3}
\expandafter\def\csname val@rev.desc.d2.uiv.p2.T.D.acc-b\endcsname{0.840}
\expandafter\def\csname cnt@rev.desc.d2.uiv.p2.T.D.acc-b\endcsname{1}
\expandafter\def\csname pct@rev.desc.d2.uiv.p2.T.D.acc-b\endcsname{84.0}
\expandafter\def\csname ppt@rev.desc.d2.uiv.p2.T.D.acc-b\endcsname{$+$84.0}
\expandafter\def\csname val@rev.desc.d2.uiv.p2.T.D.cond01\endcsname{0.250}
\expandafter\def\csname cnt@rev.desc.d2.uiv.p2.T.D.cond01\endcsname{0}
\expandafter\def\csname pct@rev.desc.d2.uiv.p2.T.D.cond01\endcsname{25.0}
\expandafter\def\csname ppt@rev.desc.d2.uiv.p2.T.D.cond01\endcsname{$+$25.0}
\expandafter\def\csname val@rev.desc.d2.uiv.p2.T.D.cond10\endcsname{0.042}
\expandafter\def\csname cnt@rev.desc.d2.uiv.p2.T.D.cond10\endcsname{0}
\expandafter\def\csname pct@rev.desc.d2.uiv.p2.T.D.cond10\endcsname{4.2}
\expandafter\def\csname ppt@rev.desc.d2.uiv.p2.T.D.cond10\endcsname{$+$4.2}
\expandafter\def\csname val@rev.desc.d2.uiv.p2.T.D.n\endcsname{144}
\expandafter\def\csname cnt@rev.desc.d2.uiv.p2.T.D.n\endcsname{144}
\expandafter\def\csname pct@rev.desc.d2.uiv.p2.T.D.n\endcsname{14400.0}
\expandafter\def\csname ppt@rev.desc.d2.uiv.p2.T.D.n\endcsname{$+$14400.0}
\expandafter\def\csname val@rev.desc.d2.uiv.p2.T.D.n-apps\endcsname{24}
\expandafter\def\csname cnt@rev.desc.d2.uiv.p2.T.D.n-apps\endcsname{24}
\expandafter\def\csname pct@rev.desc.d2.uiv.p2.T.D.n-apps\endcsname{2400.0}
\expandafter\def\csname ppt@rev.desc.d2.uiv.p2.T.D.n-apps\endcsname{$+$2400.0}
\expandafter\def\csname val@rev.desc.d2.uiv.p2.T.D.n01\endcsname{6}
\expandafter\def\csname cnt@rev.desc.d2.uiv.p2.T.D.n01\endcsname{6}
\expandafter\def\csname pct@rev.desc.d2.uiv.p2.T.D.n01\endcsname{600.0}
\expandafter\def\csname ppt@rev.desc.d2.uiv.p2.T.D.n01\endcsname{$+$600.0}
\expandafter\def\csname val@rev.desc.d2.uiv.p2.T.D.n10\endcsname{5}
\expandafter\def\csname cnt@rev.desc.d2.uiv.p2.T.D.n10\endcsname{5}
\expandafter\def\csname pct@rev.desc.d2.uiv.p2.T.D.n10\endcsname{500.0}
\expandafter\def\csname ppt@rev.desc.d2.uiv.p2.T.D.n10\endcsname{$+$500.0}
\expandafter\def\csname val@rev.desc.d2.uiv.p2.T.D.rate-any\endcsname{0.076}
\expandafter\def\csname cnt@rev.desc.d2.uiv.p2.T.D.rate-any\endcsname{0}
\expandafter\def\csname pct@rev.desc.d2.uiv.p2.T.D.rate-any\endcsname{7.6}
\expandafter\def\csname ppt@rev.desc.d2.uiv.p2.T.D.rate-any\endcsname{$+$7.6}
\expandafter\def\csname val@rev.desc.d2.uiv.p2.T.D.rate01\endcsname{0.042}
\expandafter\def\csname cnt@rev.desc.d2.uiv.p2.T.D.rate01\endcsname{0}
\expandafter\def\csname pct@rev.desc.d2.uiv.p2.T.D.rate01\endcsname{4.2}
\expandafter\def\csname ppt@rev.desc.d2.uiv.p2.T.D.rate01\endcsname{$+$4.2}
\expandafter\def\csname val@rev.desc.d2.uiv.p2.T.D.rate10\endcsname{0.035}
\expandafter\def\csname cnt@rev.desc.d2.uiv.p2.T.D.rate10\endcsname{0}
\expandafter\def\csname pct@rev.desc.d2.uiv.p2.T.D.rate10\endcsname{3.5}
\expandafter\def\csname ppt@rev.desc.d2.uiv.p2.T.D.rate10\endcsname{$+$3.5}
\expandafter\def\csname val@rev.desc.d2.uiv.p2.T.F.center-d1.acc-a\endcsname{0.778}
\expandafter\def\csname cnt@rev.desc.d2.uiv.p2.T.F.center-d1.acc-a\endcsname{1}
\expandafter\def\csname pct@rev.desc.d2.uiv.p2.T.F.center-d1.acc-a\endcsname{77.8}
\expandafter\def\csname ppt@rev.desc.d2.uiv.p2.T.F.center-d1.acc-a\endcsname{$+$77.8}
\expandafter\def\csname val@rev.desc.d2.uiv.p2.T.F.center-d1.acc-b\endcsname{0.812}
\expandafter\def\csname cnt@rev.desc.d2.uiv.p2.T.F.center-d1.acc-b\endcsname{1}
\expandafter\def\csname pct@rev.desc.d2.uiv.p2.T.F.center-d1.acc-b\endcsname{81.2}
\expandafter\def\csname ppt@rev.desc.d2.uiv.p2.T.F.center-d1.acc-b\endcsname{$+$81.2}
\expandafter\def\csname val@rev.desc.d2.uiv.p2.T.F.center-d1.cond01\endcsname{0.250}
\expandafter\def\csname cnt@rev.desc.d2.uiv.p2.T.F.center-d1.cond01\endcsname{0}
\expandafter\def\csname pct@rev.desc.d2.uiv.p2.T.F.center-d1.cond01\endcsname{25.0}
\expandafter\def\csname ppt@rev.desc.d2.uiv.p2.T.F.center-d1.cond01\endcsname{$+$25.0}
\expandafter\def\csname val@rev.desc.d2.uiv.p2.T.F.center-d1.cond10\endcsname{0.027}
\expandafter\def\csname cnt@rev.desc.d2.uiv.p2.T.F.center-d1.cond10\endcsname{0}
\expandafter\def\csname pct@rev.desc.d2.uiv.p2.T.F.center-d1.cond10\endcsname{2.7}
\expandafter\def\csname ppt@rev.desc.d2.uiv.p2.T.F.center-d1.cond10\endcsname{$+$2.7}
\expandafter\def\csname val@rev.desc.d2.uiv.p2.T.F.center-d1.n\endcsname{144}
\expandafter\def\csname cnt@rev.desc.d2.uiv.p2.T.F.center-d1.n\endcsname{144}
\expandafter\def\csname pct@rev.desc.d2.uiv.p2.T.F.center-d1.n\endcsname{14400.0}
\expandafter\def\csname ppt@rev.desc.d2.uiv.p2.T.F.center-d1.n\endcsname{$+$14400.0}
\expandafter\def\csname val@rev.desc.d2.uiv.p2.T.F.center-d1.n-apps\endcsname{24}
\expandafter\def\csname cnt@rev.desc.d2.uiv.p2.T.F.center-d1.n-apps\endcsname{24}
\expandafter\def\csname pct@rev.desc.d2.uiv.p2.T.F.center-d1.n-apps\endcsname{2400.0}
\expandafter\def\csname ppt@rev.desc.d2.uiv.p2.T.F.center-d1.n-apps\endcsname{$+$2400.0}
\expandafter\def\csname val@rev.desc.d2.uiv.p2.T.F.center-d1.n01\endcsname{8}
\expandafter\def\csname cnt@rev.desc.d2.uiv.p2.T.F.center-d1.n01\endcsname{8}
\expandafter\def\csname pct@rev.desc.d2.uiv.p2.T.F.center-d1.n01\endcsname{800.0}
\expandafter\def\csname ppt@rev.desc.d2.uiv.p2.T.F.center-d1.n01\endcsname{$+$800.0}
\expandafter\def\csname val@rev.desc.d2.uiv.p2.T.F.center-d1.n10\endcsname{3}
\expandafter\def\csname cnt@rev.desc.d2.uiv.p2.T.F.center-d1.n10\endcsname{3}
\expandafter\def\csname pct@rev.desc.d2.uiv.p2.T.F.center-d1.n10\endcsname{300.0}
\expandafter\def\csname ppt@rev.desc.d2.uiv.p2.T.F.center-d1.n10\endcsname{$+$300.0}
\expandafter\def\csname val@rev.desc.d2.uiv.p2.T.F.center-d1.rate-any\endcsname{0.076}
\expandafter\def\csname cnt@rev.desc.d2.uiv.p2.T.F.center-d1.rate-any\endcsname{0}
\expandafter\def\csname pct@rev.desc.d2.uiv.p2.T.F.center-d1.rate-any\endcsname{7.6}
\expandafter\def\csname ppt@rev.desc.d2.uiv.p2.T.F.center-d1.rate-any\endcsname{$+$7.6}
\expandafter\def\csname val@rev.desc.d2.uiv.p2.T.F.center-d1.rate01\endcsname{0.056}
\expandafter\def\csname cnt@rev.desc.d2.uiv.p2.T.F.center-d1.rate01\endcsname{0}
\expandafter\def\csname pct@rev.desc.d2.uiv.p2.T.F.center-d1.rate01\endcsname{5.6}
\expandafter\def\csname ppt@rev.desc.d2.uiv.p2.T.F.center-d1.rate01\endcsname{$+$5.6}
\expandafter\def\csname val@rev.desc.d2.uiv.p2.T.F.center-d1.rate10\endcsname{0.021}
\expandafter\def\csname cnt@rev.desc.d2.uiv.p2.T.F.center-d1.rate10\endcsname{0}
\expandafter\def\csname pct@rev.desc.d2.uiv.p2.T.F.center-d1.rate10\endcsname{2.1}
\expandafter\def\csname ppt@rev.desc.d2.uiv.p2.T.F.center-d1.rate10\endcsname{$+$2.1}
\expandafter\def\csname val@rev.desc.d2.uiv.p2.T.F.context-d1.acc-a\endcsname{0.778}
\expandafter\def\csname cnt@rev.desc.d2.uiv.p2.T.F.context-d1.acc-a\endcsname{1}
\expandafter\def\csname pct@rev.desc.d2.uiv.p2.T.F.context-d1.acc-a\endcsname{77.8}
\expandafter\def\csname ppt@rev.desc.d2.uiv.p2.T.F.context-d1.acc-a\endcsname{$+$77.8}
\expandafter\def\csname val@rev.desc.d2.uiv.p2.T.F.context-d1.acc-b\endcsname{0.840}
\expandafter\def\csname cnt@rev.desc.d2.uiv.p2.T.F.context-d1.acc-b\endcsname{1}
\expandafter\def\csname pct@rev.desc.d2.uiv.p2.T.F.context-d1.acc-b\endcsname{84.0}
\expandafter\def\csname ppt@rev.desc.d2.uiv.p2.T.F.context-d1.acc-b\endcsname{$+$84.0}
\expandafter\def\csname val@rev.desc.d2.uiv.p2.T.F.context-d1.cond01\endcsname{0.312}
\expandafter\def\csname cnt@rev.desc.d2.uiv.p2.T.F.context-d1.cond01\endcsname{0}
\expandafter\def\csname pct@rev.desc.d2.uiv.p2.T.F.context-d1.cond01\endcsname{31.2}
\expandafter\def\csname ppt@rev.desc.d2.uiv.p2.T.F.context-d1.cond01\endcsname{$+$31.2}
\expandafter\def\csname val@rev.desc.d2.uiv.p2.T.F.context-d1.cond10\endcsname{0.009}
\expandafter\def\csname cnt@rev.desc.d2.uiv.p2.T.F.context-d1.cond10\endcsname{0}
\expandafter\def\csname pct@rev.desc.d2.uiv.p2.T.F.context-d1.cond10\endcsname{0.9}
\expandafter\def\csname ppt@rev.desc.d2.uiv.p2.T.F.context-d1.cond10\endcsname{$+$0.9}
\expandafter\def\csname val@rev.desc.d2.uiv.p2.T.F.context-d1.n\endcsname{144}
\expandafter\def\csname cnt@rev.desc.d2.uiv.p2.T.F.context-d1.n\endcsname{144}
\expandafter\def\csname pct@rev.desc.d2.uiv.p2.T.F.context-d1.n\endcsname{14400.0}
\expandafter\def\csname ppt@rev.desc.d2.uiv.p2.T.F.context-d1.n\endcsname{$+$14400.0}
\expandafter\def\csname val@rev.desc.d2.uiv.p2.T.F.context-d1.n-apps\endcsname{24}
\expandafter\def\csname cnt@rev.desc.d2.uiv.p2.T.F.context-d1.n-apps\endcsname{24}
\expandafter\def\csname pct@rev.desc.d2.uiv.p2.T.F.context-d1.n-apps\endcsname{2400.0}
\expandafter\def\csname ppt@rev.desc.d2.uiv.p2.T.F.context-d1.n-apps\endcsname{$+$2400.0}
\expandafter\def\csname val@rev.desc.d2.uiv.p2.T.F.context-d1.n01\endcsname{10}
\expandafter\def\csname cnt@rev.desc.d2.uiv.p2.T.F.context-d1.n01\endcsname{10}
\expandafter\def\csname pct@rev.desc.d2.uiv.p2.T.F.context-d1.n01\endcsname{1000.0}
\expandafter\def\csname ppt@rev.desc.d2.uiv.p2.T.F.context-d1.n01\endcsname{$+$1000.0}
\expandafter\def\csname val@rev.desc.d2.uiv.p2.T.F.context-d1.n10\endcsname{1}
\expandafter\def\csname cnt@rev.desc.d2.uiv.p2.T.F.context-d1.n10\endcsname{1}
\expandafter\def\csname pct@rev.desc.d2.uiv.p2.T.F.context-d1.n10\endcsname{100.0}
\expandafter\def\csname ppt@rev.desc.d2.uiv.p2.T.F.context-d1.n10\endcsname{$+$100.0}
\expandafter\def\csname val@rev.desc.d2.uiv.p2.T.F.context-d1.rate-any\endcsname{0.076}
\expandafter\def\csname cnt@rev.desc.d2.uiv.p2.T.F.context-d1.rate-any\endcsname{0}
\expandafter\def\csname pct@rev.desc.d2.uiv.p2.T.F.context-d1.rate-any\endcsname{7.6}
\expandafter\def\csname ppt@rev.desc.d2.uiv.p2.T.F.context-d1.rate-any\endcsname{$+$7.6}
\expandafter\def\csname val@rev.desc.d2.uiv.p2.T.F.context-d1.rate01\endcsname{0.069}
\expandafter\def\csname cnt@rev.desc.d2.uiv.p2.T.F.context-d1.rate01\endcsname{0}
\expandafter\def\csname pct@rev.desc.d2.uiv.p2.T.F.context-d1.rate01\endcsname{6.9}
\expandafter\def\csname ppt@rev.desc.d2.uiv.p2.T.F.context-d1.rate01\endcsname{$+$6.9}
\expandafter\def\csname val@rev.desc.d2.uiv.p2.T.F.context-d1.rate10\endcsname{0.007}
\expandafter\def\csname cnt@rev.desc.d2.uiv.p2.T.F.context-d1.rate10\endcsname{0}
\expandafter\def\csname pct@rev.desc.d2.uiv.p2.T.F.context-d1.rate10\endcsname{0.7}
\expandafter\def\csname ppt@rev.desc.d2.uiv.p2.T.F.context-d1.rate10\endcsname{$+$0.7}
\expandafter\def\csname val@rev.desc.d2.uiv.p2.T.F.dual-d2.acc-a\endcsname{0.729}
\expandafter\def\csname cnt@rev.desc.d2.uiv.p2.T.F.dual-d2.acc-a\endcsname{1}
\expandafter\def\csname pct@rev.desc.d2.uiv.p2.T.F.dual-d2.acc-a\endcsname{72.9}
\expandafter\def\csname ppt@rev.desc.d2.uiv.p2.T.F.dual-d2.acc-a\endcsname{$+$72.9}
\expandafter\def\csname val@rev.desc.d2.uiv.p2.T.F.dual-d2.acc-b\endcsname{0.778}
\expandafter\def\csname cnt@rev.desc.d2.uiv.p2.T.F.dual-d2.acc-b\endcsname{1}
\expandafter\def\csname pct@rev.desc.d2.uiv.p2.T.F.dual-d2.acc-b\endcsname{77.8}
\expandafter\def\csname ppt@rev.desc.d2.uiv.p2.T.F.dual-d2.acc-b\endcsname{$+$77.8}
\expandafter\def\csname val@rev.desc.d2.uiv.p2.T.F.dual-d2.cond01\endcsname{0.385}
\expandafter\def\csname cnt@rev.desc.d2.uiv.p2.T.F.dual-d2.cond01\endcsname{0}
\expandafter\def\csname pct@rev.desc.d2.uiv.p2.T.F.dual-d2.cond01\endcsname{38.5}
\expandafter\def\csname ppt@rev.desc.d2.uiv.p2.T.F.dual-d2.cond01\endcsname{$+$38.5}
\expandafter\def\csname val@rev.desc.d2.uiv.p2.T.F.dual-d2.cond10\endcsname{0.076}
\expandafter\def\csname cnt@rev.desc.d2.uiv.p2.T.F.dual-d2.cond10\endcsname{0}
\expandafter\def\csname pct@rev.desc.d2.uiv.p2.T.F.dual-d2.cond10\endcsname{7.6}
\expandafter\def\csname ppt@rev.desc.d2.uiv.p2.T.F.dual-d2.cond10\endcsname{$+$7.6}
\expandafter\def\csname val@rev.desc.d2.uiv.p2.T.F.dual-d2.n\endcsname{144}
\expandafter\def\csname cnt@rev.desc.d2.uiv.p2.T.F.dual-d2.n\endcsname{144}
\expandafter\def\csname pct@rev.desc.d2.uiv.p2.T.F.dual-d2.n\endcsname{14400.0}
\expandafter\def\csname ppt@rev.desc.d2.uiv.p2.T.F.dual-d2.n\endcsname{$+$14400.0}
\expandafter\def\csname val@rev.desc.d2.uiv.p2.T.F.dual-d2.n-apps\endcsname{24}
\expandafter\def\csname cnt@rev.desc.d2.uiv.p2.T.F.dual-d2.n-apps\endcsname{24}
\expandafter\def\csname pct@rev.desc.d2.uiv.p2.T.F.dual-d2.n-apps\endcsname{2400.0}
\expandafter\def\csname ppt@rev.desc.d2.uiv.p2.T.F.dual-d2.n-apps\endcsname{$+$2400.0}
\expandafter\def\csname val@rev.desc.d2.uiv.p2.T.F.dual-d2.n01\endcsname{15}
\expandafter\def\csname cnt@rev.desc.d2.uiv.p2.T.F.dual-d2.n01\endcsname{15}
\expandafter\def\csname pct@rev.desc.d2.uiv.p2.T.F.dual-d2.n01\endcsname{1500.0}
\expandafter\def\csname ppt@rev.desc.d2.uiv.p2.T.F.dual-d2.n01\endcsname{$+$1500.0}
\expandafter\def\csname val@rev.desc.d2.uiv.p2.T.F.dual-d2.n10\endcsname{8}
\expandafter\def\csname cnt@rev.desc.d2.uiv.p2.T.F.dual-d2.n10\endcsname{8}
\expandafter\def\csname pct@rev.desc.d2.uiv.p2.T.F.dual-d2.n10\endcsname{800.0}
\expandafter\def\csname ppt@rev.desc.d2.uiv.p2.T.F.dual-d2.n10\endcsname{$+$800.0}
\expandafter\def\csname val@rev.desc.d2.uiv.p2.T.F.dual-d2.rate-any\endcsname{0.160}
\expandafter\def\csname cnt@rev.desc.d2.uiv.p2.T.F.dual-d2.rate-any\endcsname{0}
\expandafter\def\csname pct@rev.desc.d2.uiv.p2.T.F.dual-d2.rate-any\endcsname{16.0}
\expandafter\def\csname ppt@rev.desc.d2.uiv.p2.T.F.dual-d2.rate-any\endcsname{$+$16.0}
\expandafter\def\csname val@rev.desc.d2.uiv.p2.T.F.dual-d2.rate01\endcsname{0.104}
\expandafter\def\csname cnt@rev.desc.d2.uiv.p2.T.F.dual-d2.rate01\endcsname{0}
\expandafter\def\csname pct@rev.desc.d2.uiv.p2.T.F.dual-d2.rate01\endcsname{10.4}
\expandafter\def\csname ppt@rev.desc.d2.uiv.p2.T.F.dual-d2.rate01\endcsname{$+$10.4}
\expandafter\def\csname val@rev.desc.d2.uiv.p2.T.F.dual-d2.rate10\endcsname{0.056}
\expandafter\def\csname cnt@rev.desc.d2.uiv.p2.T.F.dual-d2.rate10\endcsname{0}
\expandafter\def\csname pct@rev.desc.d2.uiv.p2.T.F.dual-d2.rate10\endcsname{5.6}
\expandafter\def\csname ppt@rev.desc.d2.uiv.p2.T.F.dual-d2.rate10\endcsname{$+$5.6}
\expandafter\def\csname val@rev.desc.d2.uiv.p2.T.F.grid2-d3.acc-a\endcsname{0.757}
\expandafter\def\csname cnt@rev.desc.d2.uiv.p2.T.F.grid2-d3.acc-a\endcsname{1}
\expandafter\def\csname pct@rev.desc.d2.uiv.p2.T.F.grid2-d3.acc-a\endcsname{75.7}
\expandafter\def\csname ppt@rev.desc.d2.uiv.p2.T.F.grid2-d3.acc-a\endcsname{$+$75.7}
\expandafter\def\csname val@rev.desc.d2.uiv.p2.T.F.grid2-d3.acc-b\endcsname{0.833}
\expandafter\def\csname cnt@rev.desc.d2.uiv.p2.T.F.grid2-d3.acc-b\endcsname{1}
\expandafter\def\csname pct@rev.desc.d2.uiv.p2.T.F.grid2-d3.acc-b\endcsname{83.3}
\expandafter\def\csname ppt@rev.desc.d2.uiv.p2.T.F.grid2-d3.acc-b\endcsname{$+$83.3}
\expandafter\def\csname val@rev.desc.d2.uiv.p2.T.F.grid2-d3.cond01\endcsname{0.514}
\expandafter\def\csname cnt@rev.desc.d2.uiv.p2.T.F.grid2-d3.cond01\endcsname{1}
\expandafter\def\csname pct@rev.desc.d2.uiv.p2.T.F.grid2-d3.cond01\endcsname{51.4}
\expandafter\def\csname ppt@rev.desc.d2.uiv.p2.T.F.grid2-d3.cond01\endcsname{$+$51.4}
\expandafter\def\csname val@rev.desc.d2.uiv.p2.T.F.grid2-d3.cond10\endcsname{0.064}
\expandafter\def\csname cnt@rev.desc.d2.uiv.p2.T.F.grid2-d3.cond10\endcsname{0}
\expandafter\def\csname pct@rev.desc.d2.uiv.p2.T.F.grid2-d3.cond10\endcsname{6.4}
\expandafter\def\csname ppt@rev.desc.d2.uiv.p2.T.F.grid2-d3.cond10\endcsname{$+$6.4}
\expandafter\def\csname val@rev.desc.d2.uiv.p2.T.F.grid2-d3.n\endcsname{144}
\expandafter\def\csname cnt@rev.desc.d2.uiv.p2.T.F.grid2-d3.n\endcsname{144}
\expandafter\def\csname pct@rev.desc.d2.uiv.p2.T.F.grid2-d3.n\endcsname{14400.0}
\expandafter\def\csname ppt@rev.desc.d2.uiv.p2.T.F.grid2-d3.n\endcsname{$+$14400.0}
\expandafter\def\csname val@rev.desc.d2.uiv.p2.T.F.grid2-d3.n-apps\endcsname{24}
\expandafter\def\csname cnt@rev.desc.d2.uiv.p2.T.F.grid2-d3.n-apps\endcsname{24}
\expandafter\def\csname pct@rev.desc.d2.uiv.p2.T.F.grid2-d3.n-apps\endcsname{2400.0}
\expandafter\def\csname ppt@rev.desc.d2.uiv.p2.T.F.grid2-d3.n-apps\endcsname{$+$2400.0}
\expandafter\def\csname val@rev.desc.d2.uiv.p2.T.F.grid2-d3.n01\endcsname{18}
\expandafter\def\csname cnt@rev.desc.d2.uiv.p2.T.F.grid2-d3.n01\endcsname{18}
\expandafter\def\csname pct@rev.desc.d2.uiv.p2.T.F.grid2-d3.n01\endcsname{1800.0}
\expandafter\def\csname ppt@rev.desc.d2.uiv.p2.T.F.grid2-d3.n01\endcsname{$+$1800.0}
\expandafter\def\csname val@rev.desc.d2.uiv.p2.T.F.grid2-d3.n10\endcsname{7}
\expandafter\def\csname cnt@rev.desc.d2.uiv.p2.T.F.grid2-d3.n10\endcsname{7}
\expandafter\def\csname pct@rev.desc.d2.uiv.p2.T.F.grid2-d3.n10\endcsname{700.0}
\expandafter\def\csname ppt@rev.desc.d2.uiv.p2.T.F.grid2-d3.n10\endcsname{$+$700.0}
\expandafter\def\csname val@rev.desc.d2.uiv.p2.T.F.grid2-d3.rate-any\endcsname{0.174}
\expandafter\def\csname cnt@rev.desc.d2.uiv.p2.T.F.grid2-d3.rate-any\endcsname{0}
\expandafter\def\csname pct@rev.desc.d2.uiv.p2.T.F.grid2-d3.rate-any\endcsname{17.4}
\expandafter\def\csname ppt@rev.desc.d2.uiv.p2.T.F.grid2-d3.rate-any\endcsname{$+$17.4}
\expandafter\def\csname val@rev.desc.d2.uiv.p2.T.F.grid2-d3.rate01\endcsname{0.125}
\expandafter\def\csname cnt@rev.desc.d2.uiv.p2.T.F.grid2-d3.rate01\endcsname{0}
\expandafter\def\csname pct@rev.desc.d2.uiv.p2.T.F.grid2-d3.rate01\endcsname{12.5}
\expandafter\def\csname ppt@rev.desc.d2.uiv.p2.T.F.grid2-d3.rate01\endcsname{$+$12.5}
\expandafter\def\csname val@rev.desc.d2.uiv.p2.T.F.grid2-d3.rate10\endcsname{0.049}
\expandafter\def\csname cnt@rev.desc.d2.uiv.p2.T.F.grid2-d3.rate10\endcsname{0}
\expandafter\def\csname pct@rev.desc.d2.uiv.p2.T.F.grid2-d3.rate10\endcsname{4.9}
\expandafter\def\csname ppt@rev.desc.d2.uiv.p2.T.F.grid2-d3.rate10\endcsname{$+$4.9}
\expandafter\def\csname val@rev.desc.d2.uiv.p2.T.F.zoomclick-d3.acc-a\endcsname{0.778}
\expandafter\def\csname cnt@rev.desc.d2.uiv.p2.T.F.zoomclick-d3.acc-a\endcsname{1}
\expandafter\def\csname pct@rev.desc.d2.uiv.p2.T.F.zoomclick-d3.acc-a\endcsname{77.8}
\expandafter\def\csname ppt@rev.desc.d2.uiv.p2.T.F.zoomclick-d3.acc-a\endcsname{$+$77.8}
\expandafter\def\csname val@rev.desc.d2.uiv.p2.T.F.zoomclick-d3.acc-b\endcsname{0.833}
\expandafter\def\csname cnt@rev.desc.d2.uiv.p2.T.F.zoomclick-d3.acc-b\endcsname{1}
\expandafter\def\csname pct@rev.desc.d2.uiv.p2.T.F.zoomclick-d3.acc-b\endcsname{83.3}
\expandafter\def\csname ppt@rev.desc.d2.uiv.p2.T.F.zoomclick-d3.acc-b\endcsname{$+$83.3}
\expandafter\def\csname val@rev.desc.d2.uiv.p2.T.F.zoomclick-d3.cond01\endcsname{0.344}
\expandafter\def\csname cnt@rev.desc.d2.uiv.p2.T.F.zoomclick-d3.cond01\endcsname{0}
\expandafter\def\csname pct@rev.desc.d2.uiv.p2.T.F.zoomclick-d3.cond01\endcsname{34.4}
\expandafter\def\csname ppt@rev.desc.d2.uiv.p2.T.F.zoomclick-d3.cond01\endcsname{$+$34.4}
\expandafter\def\csname val@rev.desc.d2.uiv.p2.T.F.zoomclick-d3.cond10\endcsname{0.027}
\expandafter\def\csname cnt@rev.desc.d2.uiv.p2.T.F.zoomclick-d3.cond10\endcsname{0}
\expandafter\def\csname pct@rev.desc.d2.uiv.p2.T.F.zoomclick-d3.cond10\endcsname{2.7}
\expandafter\def\csname ppt@rev.desc.d2.uiv.p2.T.F.zoomclick-d3.cond10\endcsname{$+$2.7}
\expandafter\def\csname val@rev.desc.d2.uiv.p2.T.F.zoomclick-d3.n\endcsname{144}
\expandafter\def\csname cnt@rev.desc.d2.uiv.p2.T.F.zoomclick-d3.n\endcsname{144}
\expandafter\def\csname pct@rev.desc.d2.uiv.p2.T.F.zoomclick-d3.n\endcsname{14400.0}
\expandafter\def\csname ppt@rev.desc.d2.uiv.p2.T.F.zoomclick-d3.n\endcsname{$+$14400.0}
\expandafter\def\csname val@rev.desc.d2.uiv.p2.T.F.zoomclick-d3.n-apps\endcsname{24}
\expandafter\def\csname cnt@rev.desc.d2.uiv.p2.T.F.zoomclick-d3.n-apps\endcsname{24}
\expandafter\def\csname pct@rev.desc.d2.uiv.p2.T.F.zoomclick-d3.n-apps\endcsname{2400.0}
\expandafter\def\csname ppt@rev.desc.d2.uiv.p2.T.F.zoomclick-d3.n-apps\endcsname{$+$2400.0}
\expandafter\def\csname val@rev.desc.d2.uiv.p2.T.F.zoomclick-d3.n01\endcsname{11}
\expandafter\def\csname cnt@rev.desc.d2.uiv.p2.T.F.zoomclick-d3.n01\endcsname{11}
\expandafter\def\csname pct@rev.desc.d2.uiv.p2.T.F.zoomclick-d3.n01\endcsname{1100.0}
\expandafter\def\csname ppt@rev.desc.d2.uiv.p2.T.F.zoomclick-d3.n01\endcsname{$+$1100.0}
\expandafter\def\csname val@rev.desc.d2.uiv.p2.T.F.zoomclick-d3.n10\endcsname{3}
\expandafter\def\csname cnt@rev.desc.d2.uiv.p2.T.F.zoomclick-d3.n10\endcsname{3}
\expandafter\def\csname pct@rev.desc.d2.uiv.p2.T.F.zoomclick-d3.n10\endcsname{300.0}
\expandafter\def\csname ppt@rev.desc.d2.uiv.p2.T.F.zoomclick-d3.n10\endcsname{$+$300.0}
\expandafter\def\csname val@rev.desc.d2.uiv.p2.T.F.zoomclick-d3.rate-any\endcsname{0.097}
\expandafter\def\csname cnt@rev.desc.d2.uiv.p2.T.F.zoomclick-d3.rate-any\endcsname{0}
\expandafter\def\csname pct@rev.desc.d2.uiv.p2.T.F.zoomclick-d3.rate-any\endcsname{9.7}
\expandafter\def\csname ppt@rev.desc.d2.uiv.p2.T.F.zoomclick-d3.rate-any\endcsname{$+$9.7}
\expandafter\def\csname val@rev.desc.d2.uiv.p2.T.F.zoomclick-d3.rate01\endcsname{0.076}
\expandafter\def\csname cnt@rev.desc.d2.uiv.p2.T.F.zoomclick-d3.rate01\endcsname{0}
\expandafter\def\csname pct@rev.desc.d2.uiv.p2.T.F.zoomclick-d3.rate01\endcsname{7.6}
\expandafter\def\csname ppt@rev.desc.d2.uiv.p2.T.F.zoomclick-d3.rate01\endcsname{$+$7.6}
\expandafter\def\csname val@rev.desc.d2.uiv.p2.T.F.zoomclick-d3.rate10\endcsname{0.021}
\expandafter\def\csname cnt@rev.desc.d2.uiv.p2.T.F.zoomclick-d3.rate10\endcsname{0}
\expandafter\def\csname pct@rev.desc.d2.uiv.p2.T.F.zoomclick-d3.rate10\endcsname{2.1}
\expandafter\def\csname ppt@rev.desc.d2.uiv.p2.T.F.zoomclick-d3.rate10\endcsname{$+$2.1}
\expandafter\def\csname val@rev.desc.d2.uiv.rep.haiku.C.D.acc-a\endcsname{1.000}
\expandafter\def\csname cnt@rev.desc.d2.uiv.rep.haiku.C.D.acc-a\endcsname{1}
\expandafter\def\csname pct@rev.desc.d2.uiv.rep.haiku.C.D.acc-a\endcsname{100.0}
\expandafter\def\csname ppt@rev.desc.d2.uiv.rep.haiku.C.D.acc-a\endcsname{$+$100.0}
\expandafter\def\csname val@rev.desc.d2.uiv.rep.haiku.C.D.acc-b\endcsname{0.900}
\expandafter\def\csname cnt@rev.desc.d2.uiv.rep.haiku.C.D.acc-b\endcsname{1}
\expandafter\def\csname pct@rev.desc.d2.uiv.rep.haiku.C.D.acc-b\endcsname{90.0}
\expandafter\def\csname ppt@rev.desc.d2.uiv.rep.haiku.C.D.acc-b\endcsname{$+$90.0}
\expandafter\def\csname val@rev.desc.d2.uiv.rep.haiku.C.D.cond10\endcsname{0.100}
\expandafter\def\csname cnt@rev.desc.d2.uiv.rep.haiku.C.D.cond10\endcsname{0}
\expandafter\def\csname pct@rev.desc.d2.uiv.rep.haiku.C.D.cond10\endcsname{10.0}
\expandafter\def\csname ppt@rev.desc.d2.uiv.rep.haiku.C.D.cond10\endcsname{$+$10.0}
\expandafter\def\csname val@rev.desc.d2.uiv.rep.haiku.C.D.n\endcsname{10}
\expandafter\def\csname cnt@rev.desc.d2.uiv.rep.haiku.C.D.n\endcsname{10}
\expandafter\def\csname pct@rev.desc.d2.uiv.rep.haiku.C.D.n\endcsname{1000.0}
\expandafter\def\csname ppt@rev.desc.d2.uiv.rep.haiku.C.D.n\endcsname{$+$1000.0}
\expandafter\def\csname val@rev.desc.d2.uiv.rep.haiku.C.D.n-apps\endcsname{7}
\expandafter\def\csname cnt@rev.desc.d2.uiv.rep.haiku.C.D.n-apps\endcsname{7}
\expandafter\def\csname pct@rev.desc.d2.uiv.rep.haiku.C.D.n-apps\endcsname{700.0}
\expandafter\def\csname ppt@rev.desc.d2.uiv.rep.haiku.C.D.n-apps\endcsname{$+$700.0}
\expandafter\def\csname val@rev.desc.d2.uiv.rep.haiku.C.D.n01\endcsname{0}
\expandafter\def\csname cnt@rev.desc.d2.uiv.rep.haiku.C.D.n01\endcsname{0}
\expandafter\def\csname pct@rev.desc.d2.uiv.rep.haiku.C.D.n01\endcsname{0.0}
\expandafter\def\csname ppt@rev.desc.d2.uiv.rep.haiku.C.D.n01\endcsname{0.0}
\expandafter\def\csname val@rev.desc.d2.uiv.rep.haiku.C.D.n10\endcsname{1}
\expandafter\def\csname cnt@rev.desc.d2.uiv.rep.haiku.C.D.n10\endcsname{1}
\expandafter\def\csname pct@rev.desc.d2.uiv.rep.haiku.C.D.n10\endcsname{100.0}
\expandafter\def\csname ppt@rev.desc.d2.uiv.rep.haiku.C.D.n10\endcsname{$+$100.0}
\expandafter\def\csname val@rev.desc.d2.uiv.rep.haiku.C.D.rate-any\endcsname{0.100}
\expandafter\def\csname cnt@rev.desc.d2.uiv.rep.haiku.C.D.rate-any\endcsname{0}
\expandafter\def\csname pct@rev.desc.d2.uiv.rep.haiku.C.D.rate-any\endcsname{10.0}
\expandafter\def\csname ppt@rev.desc.d2.uiv.rep.haiku.C.D.rate-any\endcsname{$+$10.0}
\expandafter\def\csname val@rev.desc.d2.uiv.rep.haiku.C.D.rate01\endcsname{0.000}
\expandafter\def\csname cnt@rev.desc.d2.uiv.rep.haiku.C.D.rate01\endcsname{0}
\expandafter\def\csname pct@rev.desc.d2.uiv.rep.haiku.C.D.rate01\endcsname{0.0}
\expandafter\def\csname ppt@rev.desc.d2.uiv.rep.haiku.C.D.rate01\endcsname{0.0}
\expandafter\def\csname val@rev.desc.d2.uiv.rep.haiku.C.D.rate10\endcsname{0.100}
\expandafter\def\csname cnt@rev.desc.d2.uiv.rep.haiku.C.D.rate10\endcsname{0}
\expandafter\def\csname pct@rev.desc.d2.uiv.rep.haiku.C.D.rate10\endcsname{10.0}
\expandafter\def\csname ppt@rev.desc.d2.uiv.rep.haiku.C.D.rate10\endcsname{$+$10.0}
\expandafter\def\csname val@rev.desc.d2.uiv.rep.opus.C.D.acc-a\endcsname{0.684}
\expandafter\def\csname cnt@rev.desc.d2.uiv.rep.opus.C.D.acc-a\endcsname{1}
\expandafter\def\csname pct@rev.desc.d2.uiv.rep.opus.C.D.acc-a\endcsname{68.4}
\expandafter\def\csname ppt@rev.desc.d2.uiv.rep.opus.C.D.acc-a\endcsname{$+$68.4}
\expandafter\def\csname val@rev.desc.d2.uiv.rep.opus.C.D.acc-b\endcsname{0.684}
\expandafter\def\csname cnt@rev.desc.d2.uiv.rep.opus.C.D.acc-b\endcsname{1}
\expandafter\def\csname pct@rev.desc.d2.uiv.rep.opus.C.D.acc-b\endcsname{68.4}
\expandafter\def\csname ppt@rev.desc.d2.uiv.rep.opus.C.D.acc-b\endcsname{$+$68.4}
\expandafter\def\csname val@rev.desc.d2.uiv.rep.opus.C.D.cond01\endcsname{0.000}
\expandafter\def\csname cnt@rev.desc.d2.uiv.rep.opus.C.D.cond01\endcsname{0}
\expandafter\def\csname pct@rev.desc.d2.uiv.rep.opus.C.D.cond01\endcsname{0.0}
\expandafter\def\csname ppt@rev.desc.d2.uiv.rep.opus.C.D.cond01\endcsname{0.0}
\expandafter\def\csname val@rev.desc.d2.uiv.rep.opus.C.D.cond10\endcsname{0.000}
\expandafter\def\csname cnt@rev.desc.d2.uiv.rep.opus.C.D.cond10\endcsname{0}
\expandafter\def\csname pct@rev.desc.d2.uiv.rep.opus.C.D.cond10\endcsname{0.0}
\expandafter\def\csname ppt@rev.desc.d2.uiv.rep.opus.C.D.cond10\endcsname{0.0}
\expandafter\def\csname val@rev.desc.d2.uiv.rep.opus.C.D.n\endcsname{19}
\expandafter\def\csname cnt@rev.desc.d2.uiv.rep.opus.C.D.n\endcsname{19}
\expandafter\def\csname pct@rev.desc.d2.uiv.rep.opus.C.D.n\endcsname{1900.0}
\expandafter\def\csname ppt@rev.desc.d2.uiv.rep.opus.C.D.n\endcsname{$+$1900.0}
\expandafter\def\csname val@rev.desc.d2.uiv.rep.opus.C.D.n-apps\endcsname{8}
\expandafter\def\csname cnt@rev.desc.d2.uiv.rep.opus.C.D.n-apps\endcsname{8}
\expandafter\def\csname pct@rev.desc.d2.uiv.rep.opus.C.D.n-apps\endcsname{800.0}
\expandafter\def\csname ppt@rev.desc.d2.uiv.rep.opus.C.D.n-apps\endcsname{$+$800.0}
\expandafter\def\csname val@rev.desc.d2.uiv.rep.opus.C.D.n01\endcsname{0}
\expandafter\def\csname cnt@rev.desc.d2.uiv.rep.opus.C.D.n01\endcsname{0}
\expandafter\def\csname pct@rev.desc.d2.uiv.rep.opus.C.D.n01\endcsname{0.0}
\expandafter\def\csname ppt@rev.desc.d2.uiv.rep.opus.C.D.n01\endcsname{0.0}
\expandafter\def\csname val@rev.desc.d2.uiv.rep.opus.C.D.n10\endcsname{0}
\expandafter\def\csname cnt@rev.desc.d2.uiv.rep.opus.C.D.n10\endcsname{0}
\expandafter\def\csname pct@rev.desc.d2.uiv.rep.opus.C.D.n10\endcsname{0.0}
\expandafter\def\csname ppt@rev.desc.d2.uiv.rep.opus.C.D.n10\endcsname{0.0}
\expandafter\def\csname val@rev.desc.d2.uiv.rep.opus.C.D.rate-any\endcsname{0.000}
\expandafter\def\csname cnt@rev.desc.d2.uiv.rep.opus.C.D.rate-any\endcsname{0}
\expandafter\def\csname pct@rev.desc.d2.uiv.rep.opus.C.D.rate-any\endcsname{0.0}
\expandafter\def\csname ppt@rev.desc.d2.uiv.rep.opus.C.D.rate-any\endcsname{0.0}
\expandafter\def\csname val@rev.desc.d2.uiv.rep.opus.C.D.rate01\endcsname{0.000}
\expandafter\def\csname cnt@rev.desc.d2.uiv.rep.opus.C.D.rate01\endcsname{0}
\expandafter\def\csname pct@rev.desc.d2.uiv.rep.opus.C.D.rate01\endcsname{0.0}
\expandafter\def\csname ppt@rev.desc.d2.uiv.rep.opus.C.D.rate01\endcsname{0.0}
\expandafter\def\csname val@rev.desc.d2.uiv.rep.opus.C.D.rate10\endcsname{0.000}
\expandafter\def\csname cnt@rev.desc.d2.uiv.rep.opus.C.D.rate10\endcsname{0}
\expandafter\def\csname pct@rev.desc.d2.uiv.rep.opus.C.D.rate10\endcsname{0.0}
\expandafter\def\csname ppt@rev.desc.d2.uiv.rep.opus.C.D.rate10\endcsname{0.0}
\expandafter\def\csname val@rev.desc.d2.uiv.rep.opus.T.D.acc-a\endcsname{0.840}
\expandafter\def\csname cnt@rev.desc.d2.uiv.rep.opus.T.D.acc-a\endcsname{1}
\expandafter\def\csname pct@rev.desc.d2.uiv.rep.opus.T.D.acc-a\endcsname{84.0}
\expandafter\def\csname ppt@rev.desc.d2.uiv.rep.opus.T.D.acc-a\endcsname{$+$84.0}
\expandafter\def\csname val@rev.desc.d2.uiv.rep.opus.T.D.acc-b\endcsname{0.833}
\expandafter\def\csname cnt@rev.desc.d2.uiv.rep.opus.T.D.acc-b\endcsname{1}
\expandafter\def\csname pct@rev.desc.d2.uiv.rep.opus.T.D.acc-b\endcsname{83.3}
\expandafter\def\csname ppt@rev.desc.d2.uiv.rep.opus.T.D.acc-b\endcsname{$+$83.3}
\expandafter\def\csname val@rev.desc.d2.uiv.rep.opus.T.D.cond01\endcsname{0.043}
\expandafter\def\csname cnt@rev.desc.d2.uiv.rep.opus.T.D.cond01\endcsname{0}
\expandafter\def\csname pct@rev.desc.d2.uiv.rep.opus.T.D.cond01\endcsname{4.3}
\expandafter\def\csname ppt@rev.desc.d2.uiv.rep.opus.T.D.cond01\endcsname{$+$4.3}
\expandafter\def\csname val@rev.desc.d2.uiv.rep.opus.T.D.cond10\endcsname{0.017}
\expandafter\def\csname cnt@rev.desc.d2.uiv.rep.opus.T.D.cond10\endcsname{0}
\expandafter\def\csname pct@rev.desc.d2.uiv.rep.opus.T.D.cond10\endcsname{1.7}
\expandafter\def\csname ppt@rev.desc.d2.uiv.rep.opus.T.D.cond10\endcsname{$+$1.7}
\expandafter\def\csname val@rev.desc.d2.uiv.rep.opus.T.D.n\endcsname{144}
\expandafter\def\csname cnt@rev.desc.d2.uiv.rep.opus.T.D.n\endcsname{144}
\expandafter\def\csname pct@rev.desc.d2.uiv.rep.opus.T.D.n\endcsname{14400.0}
\expandafter\def\csname ppt@rev.desc.d2.uiv.rep.opus.T.D.n\endcsname{$+$14400.0}
\expandafter\def\csname val@rev.desc.d2.uiv.rep.opus.T.D.n-apps\endcsname{24}
\expandafter\def\csname cnt@rev.desc.d2.uiv.rep.opus.T.D.n-apps\endcsname{24}
\expandafter\def\csname pct@rev.desc.d2.uiv.rep.opus.T.D.n-apps\endcsname{2400.0}
\expandafter\def\csname ppt@rev.desc.d2.uiv.rep.opus.T.D.n-apps\endcsname{$+$2400.0}
\expandafter\def\csname val@rev.desc.d2.uiv.rep.opus.T.D.n01\endcsname{1}
\expandafter\def\csname cnt@rev.desc.d2.uiv.rep.opus.T.D.n01\endcsname{1}
\expandafter\def\csname pct@rev.desc.d2.uiv.rep.opus.T.D.n01\endcsname{100.0}
\expandafter\def\csname ppt@rev.desc.d2.uiv.rep.opus.T.D.n01\endcsname{$+$100.0}
\expandafter\def\csname val@rev.desc.d2.uiv.rep.opus.T.D.n10\endcsname{2}
\expandafter\def\csname cnt@rev.desc.d2.uiv.rep.opus.T.D.n10\endcsname{2}
\expandafter\def\csname pct@rev.desc.d2.uiv.rep.opus.T.D.n10\endcsname{200.0}
\expandafter\def\csname ppt@rev.desc.d2.uiv.rep.opus.T.D.n10\endcsname{$+$200.0}
\expandafter\def\csname val@rev.desc.d2.uiv.rep.opus.T.D.rate-any\endcsname{0.021}
\expandafter\def\csname cnt@rev.desc.d2.uiv.rep.opus.T.D.rate-any\endcsname{0}
\expandafter\def\csname pct@rev.desc.d2.uiv.rep.opus.T.D.rate-any\endcsname{2.1}
\expandafter\def\csname ppt@rev.desc.d2.uiv.rep.opus.T.D.rate-any\endcsname{$+$2.1}
\expandafter\def\csname val@rev.desc.d2.uiv.rep.opus.T.D.rate01\endcsname{0.007}
\expandafter\def\csname cnt@rev.desc.d2.uiv.rep.opus.T.D.rate01\endcsname{0}
\expandafter\def\csname pct@rev.desc.d2.uiv.rep.opus.T.D.rate01\endcsname{0.7}
\expandafter\def\csname ppt@rev.desc.d2.uiv.rep.opus.T.D.rate01\endcsname{$+$0.7}
\expandafter\def\csname val@rev.desc.d2.uiv.rep.opus.T.D.rate10\endcsname{0.014}
\expandafter\def\csname cnt@rev.desc.d2.uiv.rep.opus.T.D.rate10\endcsname{0}
\expandafter\def\csname pct@rev.desc.d2.uiv.rep.opus.T.D.rate10\endcsname{1.4}
\expandafter\def\csname ppt@rev.desc.d2.uiv.rep.opus.T.D.rate10\endcsname{$+$1.4}
\expandafter\def\csname val@rev.desc.d2.uiv.rep.sonnet.C.D.acc-a\endcsname{0.714}
\expandafter\def\csname cnt@rev.desc.d2.uiv.rep.sonnet.C.D.acc-a\endcsname{1}
\expandafter\def\csname pct@rev.desc.d2.uiv.rep.sonnet.C.D.acc-a\endcsname{71.4}
\expandafter\def\csname ppt@rev.desc.d2.uiv.rep.sonnet.C.D.acc-a\endcsname{$+$71.4}
\expandafter\def\csname val@rev.desc.d2.uiv.rep.sonnet.C.D.acc-b\endcsname{0.786}
\expandafter\def\csname cnt@rev.desc.d2.uiv.rep.sonnet.C.D.acc-b\endcsname{1}
\expandafter\def\csname pct@rev.desc.d2.uiv.rep.sonnet.C.D.acc-b\endcsname{78.6}
\expandafter\def\csname ppt@rev.desc.d2.uiv.rep.sonnet.C.D.acc-b\endcsname{$+$78.6}
\expandafter\def\csname val@rev.desc.d2.uiv.rep.sonnet.C.D.cond01\endcsname{0.500}
\expandafter\def\csname cnt@rev.desc.d2.uiv.rep.sonnet.C.D.cond01\endcsname{0}
\expandafter\def\csname pct@rev.desc.d2.uiv.rep.sonnet.C.D.cond01\endcsname{50.0}
\expandafter\def\csname ppt@rev.desc.d2.uiv.rep.sonnet.C.D.cond01\endcsname{$+$50.0}
\expandafter\def\csname val@rev.desc.d2.uiv.rep.sonnet.C.D.cond10\endcsname{0.100}
\expandafter\def\csname cnt@rev.desc.d2.uiv.rep.sonnet.C.D.cond10\endcsname{0}
\expandafter\def\csname pct@rev.desc.d2.uiv.rep.sonnet.C.D.cond10\endcsname{10.0}
\expandafter\def\csname ppt@rev.desc.d2.uiv.rep.sonnet.C.D.cond10\endcsname{$+$10.0}
\expandafter\def\csname val@rev.desc.d2.uiv.rep.sonnet.C.D.n\endcsname{14}
\expandafter\def\csname cnt@rev.desc.d2.uiv.rep.sonnet.C.D.n\endcsname{14}
\expandafter\def\csname pct@rev.desc.d2.uiv.rep.sonnet.C.D.n\endcsname{1400.0}
\expandafter\def\csname ppt@rev.desc.d2.uiv.rep.sonnet.C.D.n\endcsname{$+$1400.0}
\expandafter\def\csname val@rev.desc.d2.uiv.rep.sonnet.C.D.n-apps\endcsname{10}
\expandafter\def\csname cnt@rev.desc.d2.uiv.rep.sonnet.C.D.n-apps\endcsname{10}
\expandafter\def\csname pct@rev.desc.d2.uiv.rep.sonnet.C.D.n-apps\endcsname{1000.0}
\expandafter\def\csname ppt@rev.desc.d2.uiv.rep.sonnet.C.D.n-apps\endcsname{$+$1000.0}
\expandafter\def\csname val@rev.desc.d2.uiv.rep.sonnet.C.D.n01\endcsname{2}
\expandafter\def\csname cnt@rev.desc.d2.uiv.rep.sonnet.C.D.n01\endcsname{2}
\expandafter\def\csname pct@rev.desc.d2.uiv.rep.sonnet.C.D.n01\endcsname{200.0}
\expandafter\def\csname ppt@rev.desc.d2.uiv.rep.sonnet.C.D.n01\endcsname{$+$200.0}
\expandafter\def\csname val@rev.desc.d2.uiv.rep.sonnet.C.D.n10\endcsname{1}
\expandafter\def\csname cnt@rev.desc.d2.uiv.rep.sonnet.C.D.n10\endcsname{1}
\expandafter\def\csname pct@rev.desc.d2.uiv.rep.sonnet.C.D.n10\endcsname{100.0}
\expandafter\def\csname ppt@rev.desc.d2.uiv.rep.sonnet.C.D.n10\endcsname{$+$100.0}
\expandafter\def\csname val@rev.desc.d2.uiv.rep.sonnet.C.D.rate-any\endcsname{0.214}
\expandafter\def\csname cnt@rev.desc.d2.uiv.rep.sonnet.C.D.rate-any\endcsname{0}
\expandafter\def\csname pct@rev.desc.d2.uiv.rep.sonnet.C.D.rate-any\endcsname{21.4}
\expandafter\def\csname ppt@rev.desc.d2.uiv.rep.sonnet.C.D.rate-any\endcsname{$+$21.4}
\expandafter\def\csname val@rev.desc.d2.uiv.rep.sonnet.C.D.rate01\endcsname{0.143}
\expandafter\def\csname cnt@rev.desc.d2.uiv.rep.sonnet.C.D.rate01\endcsname{0}
\expandafter\def\csname pct@rev.desc.d2.uiv.rep.sonnet.C.D.rate01\endcsname{14.3}
\expandafter\def\csname ppt@rev.desc.d2.uiv.rep.sonnet.C.D.rate01\endcsname{$+$14.3}
\expandafter\def\csname val@rev.desc.d2.uiv.rep.sonnet.C.D.rate10\endcsname{0.071}
\expandafter\def\csname cnt@rev.desc.d2.uiv.rep.sonnet.C.D.rate10\endcsname{0}
\expandafter\def\csname pct@rev.desc.d2.uiv.rep.sonnet.C.D.rate10\endcsname{7.1}
\expandafter\def\csname ppt@rev.desc.d2.uiv.rep.sonnet.C.D.rate10\endcsname{$+$7.1}
\expandafter\def\csname val@rev.desc.d2.uiv.rep.sonnet.T.D.acc-a\endcsname{0.833}
\expandafter\def\csname cnt@rev.desc.d2.uiv.rep.sonnet.T.D.acc-a\endcsname{1}
\expandafter\def\csname pct@rev.desc.d2.uiv.rep.sonnet.T.D.acc-a\endcsname{83.3}
\expandafter\def\csname ppt@rev.desc.d2.uiv.rep.sonnet.T.D.acc-a\endcsname{$+$83.3}
\expandafter\def\csname val@rev.desc.d2.uiv.rep.sonnet.T.D.acc-b\endcsname{0.792}
\expandafter\def\csname cnt@rev.desc.d2.uiv.rep.sonnet.T.D.acc-b\endcsname{1}
\expandafter\def\csname pct@rev.desc.d2.uiv.rep.sonnet.T.D.acc-b\endcsname{79.2}
\expandafter\def\csname ppt@rev.desc.d2.uiv.rep.sonnet.T.D.acc-b\endcsname{$+$79.2}
\expandafter\def\csname val@rev.desc.d2.uiv.rep.sonnet.T.D.cond01\endcsname{0.042}
\expandafter\def\csname cnt@rev.desc.d2.uiv.rep.sonnet.T.D.cond01\endcsname{0}
\expandafter\def\csname pct@rev.desc.d2.uiv.rep.sonnet.T.D.cond01\endcsname{4.2}
\expandafter\def\csname ppt@rev.desc.d2.uiv.rep.sonnet.T.D.cond01\endcsname{$+$4.2}
\expandafter\def\csname val@rev.desc.d2.uiv.rep.sonnet.T.D.cond10\endcsname{0.058}
\expandafter\def\csname cnt@rev.desc.d2.uiv.rep.sonnet.T.D.cond10\endcsname{0}
\expandafter\def\csname pct@rev.desc.d2.uiv.rep.sonnet.T.D.cond10\endcsname{5.8}
\expandafter\def\csname ppt@rev.desc.d2.uiv.rep.sonnet.T.D.cond10\endcsname{$+$5.8}
\expandafter\def\csname val@rev.desc.d2.uiv.rep.sonnet.T.D.n\endcsname{144}
\expandafter\def\csname cnt@rev.desc.d2.uiv.rep.sonnet.T.D.n\endcsname{144}
\expandafter\def\csname pct@rev.desc.d2.uiv.rep.sonnet.T.D.n\endcsname{14400.0}
\expandafter\def\csname ppt@rev.desc.d2.uiv.rep.sonnet.T.D.n\endcsname{$+$14400.0}
\expandafter\def\csname val@rev.desc.d2.uiv.rep.sonnet.T.D.n-apps\endcsname{24}
\expandafter\def\csname cnt@rev.desc.d2.uiv.rep.sonnet.T.D.n-apps\endcsname{24}
\expandafter\def\csname pct@rev.desc.d2.uiv.rep.sonnet.T.D.n-apps\endcsname{2400.0}
\expandafter\def\csname ppt@rev.desc.d2.uiv.rep.sonnet.T.D.n-apps\endcsname{$+$2400.0}
\expandafter\def\csname val@rev.desc.d2.uiv.rep.sonnet.T.D.n01\endcsname{1}
\expandafter\def\csname cnt@rev.desc.d2.uiv.rep.sonnet.T.D.n01\endcsname{1}
\expandafter\def\csname pct@rev.desc.d2.uiv.rep.sonnet.T.D.n01\endcsname{100.0}
\expandafter\def\csname ppt@rev.desc.d2.uiv.rep.sonnet.T.D.n01\endcsname{$+$100.0}
\expandafter\def\csname val@rev.desc.d2.uiv.rep.sonnet.T.D.n10\endcsname{7}
\expandafter\def\csname cnt@rev.desc.d2.uiv.rep.sonnet.T.D.n10\endcsname{7}
\expandafter\def\csname pct@rev.desc.d2.uiv.rep.sonnet.T.D.n10\endcsname{700.0}
\expandafter\def\csname ppt@rev.desc.d2.uiv.rep.sonnet.T.D.n10\endcsname{$+$700.0}
\expandafter\def\csname val@rev.desc.d2.uiv.rep.sonnet.T.D.rate-any\endcsname{0.056}
\expandafter\def\csname cnt@rev.desc.d2.uiv.rep.sonnet.T.D.rate-any\endcsname{0}
\expandafter\def\csname pct@rev.desc.d2.uiv.rep.sonnet.T.D.rate-any\endcsname{5.6}
\expandafter\def\csname ppt@rev.desc.d2.uiv.rep.sonnet.T.D.rate-any\endcsname{$+$5.6}
\expandafter\def\csname val@rev.desc.d2.uiv.rep.sonnet.T.D.rate01\endcsname{0.007}
\expandafter\def\csname cnt@rev.desc.d2.uiv.rep.sonnet.T.D.rate01\endcsname{0}
\expandafter\def\csname pct@rev.desc.d2.uiv.rep.sonnet.T.D.rate01\endcsname{0.7}
\expandafter\def\csname ppt@rev.desc.d2.uiv.rep.sonnet.T.D.rate01\endcsname{$+$0.7}
\expandafter\def\csname val@rev.desc.d2.uiv.rep.sonnet.T.D.rate10\endcsname{0.049}
\expandafter\def\csname cnt@rev.desc.d2.uiv.rep.sonnet.T.D.rate10\endcsname{0}
\expandafter\def\csname pct@rev.desc.d2.uiv.rep.sonnet.T.D.rate10\endcsname{4.9}
\expandafter\def\csname ppt@rev.desc.d2.uiv.rep.sonnet.T.D.rate10\endcsname{$+$4.9}
\expandafter\def\csname val@rev.desc.d3.sspro.p1.C.catalog-pooled.N-agree-stable\endcsname{146}
\expandafter\def\csname cnt@rev.desc.d3.sspro.p1.C.catalog-pooled.N-agree-stable\endcsname{146}
\expandafter\def\csname pct@rev.desc.d3.sspro.p1.C.catalog-pooled.N-agree-stable\endcsname{14600.0}
\expandafter\def\csname ppt@rev.desc.d3.sspro.p1.C.catalog-pooled.N-agree-stable\endcsname{$+$14600.0}
\expandafter\def\csname val@rev.desc.d3.sspro.p1.C.catalog-pooled.N-disagree\endcsname{33}
\expandafter\def\csname cnt@rev.desc.d3.sspro.p1.C.catalog-pooled.N-disagree\endcsname{33}
\expandafter\def\csname pct@rev.desc.d3.sspro.p1.C.catalog-pooled.N-disagree\endcsname{3300.0}
\expandafter\def\csname ppt@rev.desc.d3.sspro.p1.C.catalog-pooled.N-disagree\endcsname{$+$3300.0}
\expandafter\def\csname val@rev.desc.d3.sspro.p1.C.catalog-pooled.boot-resid.ci95.0\endcsname{-0.015}
\expandafter\def\csname cnt@rev.desc.d3.sspro.p1.C.catalog-pooled.boot-resid.ci95.0\endcsname{0}
\expandafter\def\csname pct@rev.desc.d3.sspro.p1.C.catalog-pooled.boot-resid.ci95.0\endcsname{$-$1.5}
\expandafter\def\csname ppt@rev.desc.d3.sspro.p1.C.catalog-pooled.boot-resid.ci95.0\endcsname{$-$1.5}
\expandafter\def\csname val@rev.desc.d3.sspro.p1.C.catalog-pooled.boot-resid.ci95.1\endcsname{0.045}
\expandafter\def\csname cnt@rev.desc.d3.sspro.p1.C.catalog-pooled.boot-resid.ci95.1\endcsname{0}
\expandafter\def\csname pct@rev.desc.d3.sspro.p1.C.catalog-pooled.boot-resid.ci95.1\endcsname{4.5}
\expandafter\def\csname ppt@rev.desc.d3.sspro.p1.C.catalog-pooled.boot-resid.ci95.1\endcsname{$+$4.5}
\expandafter\def\csname val@rev.desc.d3.sspro.p1.C.catalog-pooled.boot-resid.n-apps\endcsname{6}
\expandafter\def\csname cnt@rev.desc.d3.sspro.p1.C.catalog-pooled.boot-resid.n-apps\endcsname{6}
\expandafter\def\csname pct@rev.desc.d3.sspro.p1.C.catalog-pooled.boot-resid.n-apps\endcsname{600.0}
\expandafter\def\csname ppt@rev.desc.d3.sspro.p1.C.catalog-pooled.boot-resid.n-apps\endcsname{$+$600.0}
\expandafter\def\csname val@rev.desc.d3.sspro.p1.C.catalog-pooled.boot-resid.n-boot-valid\endcsname{4000}
\expandafter\def\csname cnt@rev.desc.d3.sspro.p1.C.catalog-pooled.boot-resid.n-boot-valid\endcsname{4{,}000}
\expandafter\def\csname pct@rev.desc.d3.sspro.p1.C.catalog-pooled.boot-resid.n-boot-valid\endcsname{400000.0}
\expandafter\def\csname ppt@rev.desc.d3.sspro.p1.C.catalog-pooled.boot-resid.n-boot-valid\endcsname{$+$400000.0}
\expandafter\def\csname val@rev.desc.d3.sspro.p1.C.catalog-pooled.boot-resid.value\endcsname{0.014}
\expandafter\def\csname cnt@rev.desc.d3.sspro.p1.C.catalog-pooled.boot-resid.value\endcsname{0}
\expandafter\def\csname pct@rev.desc.d3.sspro.p1.C.catalog-pooled.boot-resid.value\endcsname{1.4}
\expandafter\def\csname ppt@rev.desc.d3.sspro.p1.C.catalog-pooled.boot-resid.value\endcsname{$+$1.4}
\expandafter\def\csname val@rev.desc.d3.sspro.p1.C.catalog-pooled.boot-share.ci95.0\endcsname{0.594}
\expandafter\def\csname cnt@rev.desc.d3.sspro.p1.C.catalog-pooled.boot-share.ci95.0\endcsname{1}
\expandafter\def\csname pct@rev.desc.d3.sspro.p1.C.catalog-pooled.boot-share.ci95.0\endcsname{59.4}
\expandafter\def\csname ppt@rev.desc.d3.sspro.p1.C.catalog-pooled.boot-share.ci95.0\endcsname{$+$59.4}
\expandafter\def\csname val@rev.desc.d3.sspro.p1.C.catalog-pooled.boot-share.ci95.1\endcsname{0.885}
\expandafter\def\csname cnt@rev.desc.d3.sspro.p1.C.catalog-pooled.boot-share.ci95.1\endcsname{1}
\expandafter\def\csname pct@rev.desc.d3.sspro.p1.C.catalog-pooled.boot-share.ci95.1\endcsname{88.5}
\expandafter\def\csname ppt@rev.desc.d3.sspro.p1.C.catalog-pooled.boot-share.ci95.1\endcsname{$+$88.5}
\expandafter\def\csname val@rev.desc.d3.sspro.p1.C.catalog-pooled.boot-share.n-apps\endcsname{6}
\expandafter\def\csname cnt@rev.desc.d3.sspro.p1.C.catalog-pooled.boot-share.n-apps\endcsname{6}
\expandafter\def\csname pct@rev.desc.d3.sspro.p1.C.catalog-pooled.boot-share.n-apps\endcsname{600.0}
\expandafter\def\csname ppt@rev.desc.d3.sspro.p1.C.catalog-pooled.boot-share.n-apps\endcsname{$+$600.0}
\expandafter\def\csname val@rev.desc.d3.sspro.p1.C.catalog-pooled.boot-share.n-boot-valid\endcsname{4000}
\expandafter\def\csname cnt@rev.desc.d3.sspro.p1.C.catalog-pooled.boot-share.n-boot-valid\endcsname{4{,}000}
\expandafter\def\csname pct@rev.desc.d3.sspro.p1.C.catalog-pooled.boot-share.n-boot-valid\endcsname{400000.0}
\expandafter\def\csname ppt@rev.desc.d3.sspro.p1.C.catalog-pooled.boot-share.n-boot-valid\endcsname{$+$400000.0}
\expandafter\def\csname val@rev.desc.d3.sspro.p1.C.catalog-pooled.boot-share.value\endcsname{0.760}
\expandafter\def\csname cnt@rev.desc.d3.sspro.p1.C.catalog-pooled.boot-share.value\endcsname{1}
\expandafter\def\csname pct@rev.desc.d3.sspro.p1.C.catalog-pooled.boot-share.value\endcsname{76.0}
\expandafter\def\csname ppt@rev.desc.d3.sspro.p1.C.catalog-pooled.boot-share.value\endcsname{$+$76.0}
\expandafter\def\csname val@rev.desc.d3.sspro.p1.C.catalog-pooled.gap\endcsname{0.017}
\expandafter\def\csname cnt@rev.desc.d3.sspro.p1.C.catalog-pooled.gap\endcsname{0}
\expandafter\def\csname pct@rev.desc.d3.sspro.p1.C.catalog-pooled.gap\endcsname{1.7}
\expandafter\def\csname ppt@rev.desc.d3.sspro.p1.C.catalog-pooled.gap\endcsname{$+$1.7}
\expandafter\def\csname val@rev.desc.d3.sspro.p1.C.catalog-pooled.n\endcsname{192}
\expandafter\def\csname cnt@rev.desc.d3.sspro.p1.C.catalog-pooled.n\endcsname{192}
\expandafter\def\csname pct@rev.desc.d3.sspro.p1.C.catalog-pooled.n\endcsname{19200.0}
\expandafter\def\csname ppt@rev.desc.d3.sspro.p1.C.catalog-pooled.n\endcsname{$+$19200.0}
\expandafter\def\csname val@rev.desc.d3.sspro.p1.C.catalog-pooled.reproduces-frozen\endcsname{1}
\expandafter\def\csname cnt@rev.desc.d3.sspro.p1.C.catalog-pooled.reproduces-frozen\endcsname{1}
\expandafter\def\csname pct@rev.desc.d3.sspro.p1.C.catalog-pooled.reproduces-frozen\endcsname{100.0}
\expandafter\def\csname ppt@rev.desc.d3.sspro.p1.C.catalog-pooled.reproduces-frozen\endcsname{$+$100.0}
\expandafter\def\csname val@rev.desc.d3.sspro.p1.C.catalog-pooled.resid-agree-stable\endcsname{0.014}
\expandafter\def\csname cnt@rev.desc.d3.sspro.p1.C.catalog-pooled.resid-agree-stable\endcsname{0}
\expandafter\def\csname pct@rev.desc.d3.sspro.p1.C.catalog-pooled.resid-agree-stable\endcsname{1.4}
\expandafter\def\csname ppt@rev.desc.d3.sspro.p1.C.catalog-pooled.resid-agree-stable\endcsname{$+$1.4}
\expandafter\def\csname val@rev.desc.d3.sspro.p1.C.catalog-pooled.resid-disagree\endcsname{0.030}
\expandafter\def\csname cnt@rev.desc.d3.sspro.p1.C.catalog-pooled.resid-disagree\endcsname{0}
\expandafter\def\csname pct@rev.desc.d3.sspro.p1.C.catalog-pooled.resid-disagree\endcsname{3.0}
\expandafter\def\csname ppt@rev.desc.d3.sspro.p1.C.catalog-pooled.resid-disagree\endcsname{$+$3.0}
\expandafter\def\csname val@rev.desc.d3.sspro.p1.C.catalog-pooled.share-agree-stable\endcsname{0.760}
\expandafter\def\csname cnt@rev.desc.d3.sspro.p1.C.catalog-pooled.share-agree-stable\endcsname{1}
\expandafter\def\csname pct@rev.desc.d3.sspro.p1.C.catalog-pooled.share-agree-stable\endcsname{76.0}
\expandafter\def\csname ppt@rev.desc.d3.sspro.p1.C.catalog-pooled.share-agree-stable\endcsname{$+$76.0}
\expandafter\def\csname val@rev.desc.d3.sspro.p1.C.center-d1.N-agree-stable\endcsname{38}
\expandafter\def\csname cnt@rev.desc.d3.sspro.p1.C.center-d1.N-agree-stable\endcsname{38}
\expandafter\def\csname pct@rev.desc.d3.sspro.p1.C.center-d1.N-agree-stable\endcsname{3800.0}
\expandafter\def\csname ppt@rev.desc.d3.sspro.p1.C.center-d1.N-agree-stable\endcsname{$+$3800.0}
\expandafter\def\csname val@rev.desc.d3.sspro.p1.C.center-d1.N-disagree\endcsname{8}
\expandafter\def\csname cnt@rev.desc.d3.sspro.p1.C.center-d1.N-disagree\endcsname{8}
\expandafter\def\csname pct@rev.desc.d3.sspro.p1.C.center-d1.N-disagree\endcsname{800.0}
\expandafter\def\csname ppt@rev.desc.d3.sspro.p1.C.center-d1.N-disagree\endcsname{$+$800.0}
\expandafter\def\csname val@rev.desc.d3.sspro.p1.C.center-d1.boot-resid.ci95.0\endcsname{0.000}
\expandafter\def\csname cnt@rev.desc.d3.sspro.p1.C.center-d1.boot-resid.ci95.0\endcsname{0}
\expandafter\def\csname pct@rev.desc.d3.sspro.p1.C.center-d1.boot-resid.ci95.0\endcsname{0.0}
\expandafter\def\csname ppt@rev.desc.d3.sspro.p1.C.center-d1.boot-resid.ci95.0\endcsname{0.0}
\expandafter\def\csname val@rev.desc.d3.sspro.p1.C.center-d1.boot-resid.ci95.1\endcsname{0.079}
\expandafter\def\csname cnt@rev.desc.d3.sspro.p1.C.center-d1.boot-resid.ci95.1\endcsname{0}
\expandafter\def\csname pct@rev.desc.d3.sspro.p1.C.center-d1.boot-resid.ci95.1\endcsname{7.9}
\expandafter\def\csname ppt@rev.desc.d3.sspro.p1.C.center-d1.boot-resid.ci95.1\endcsname{$+$7.9}
\expandafter\def\csname val@rev.desc.d3.sspro.p1.C.center-d1.boot-resid.n-apps\endcsname{6}
\expandafter\def\csname cnt@rev.desc.d3.sspro.p1.C.center-d1.boot-resid.n-apps\endcsname{6}
\expandafter\def\csname pct@rev.desc.d3.sspro.p1.C.center-d1.boot-resid.n-apps\endcsname{600.0}
\expandafter\def\csname ppt@rev.desc.d3.sspro.p1.C.center-d1.boot-resid.n-apps\endcsname{$+$600.0}
\expandafter\def\csname val@rev.desc.d3.sspro.p1.C.center-d1.boot-resid.n-boot-valid\endcsname{4000}
\expandafter\def\csname cnt@rev.desc.d3.sspro.p1.C.center-d1.boot-resid.n-boot-valid\endcsname{4{,}000}
\expandafter\def\csname pct@rev.desc.d3.sspro.p1.C.center-d1.boot-resid.n-boot-valid\endcsname{400000.0}
\expandafter\def\csname ppt@rev.desc.d3.sspro.p1.C.center-d1.boot-resid.n-boot-valid\endcsname{$+$400000.0}
\expandafter\def\csname val@rev.desc.d3.sspro.p1.C.center-d1.boot-resid.value\endcsname{0.026}
\expandafter\def\csname cnt@rev.desc.d3.sspro.p1.C.center-d1.boot-resid.value\endcsname{0}
\expandafter\def\csname pct@rev.desc.d3.sspro.p1.C.center-d1.boot-resid.value\endcsname{2.6}
\expandafter\def\csname ppt@rev.desc.d3.sspro.p1.C.center-d1.boot-resid.value\endcsname{$+$2.6}
\expandafter\def\csname val@rev.desc.d3.sspro.p1.C.center-d1.boot-share.ci95.0\endcsname{0.604}
\expandafter\def\csname cnt@rev.desc.d3.sspro.p1.C.center-d1.boot-share.ci95.0\endcsname{1}
\expandafter\def\csname pct@rev.desc.d3.sspro.p1.C.center-d1.boot-share.ci95.0\endcsname{60.4}
\expandafter\def\csname ppt@rev.desc.d3.sspro.p1.C.center-d1.boot-share.ci95.0\endcsname{$+$60.4}
\expandafter\def\csname val@rev.desc.d3.sspro.p1.C.center-d1.boot-share.ci95.1\endcsname{0.917}
\expandafter\def\csname cnt@rev.desc.d3.sspro.p1.C.center-d1.boot-share.ci95.1\endcsname{1}
\expandafter\def\csname pct@rev.desc.d3.sspro.p1.C.center-d1.boot-share.ci95.1\endcsname{91.7}
\expandafter\def\csname ppt@rev.desc.d3.sspro.p1.C.center-d1.boot-share.ci95.1\endcsname{$+$91.7}
\expandafter\def\csname val@rev.desc.d3.sspro.p1.C.center-d1.boot-share.n-apps\endcsname{6}
\expandafter\def\csname cnt@rev.desc.d3.sspro.p1.C.center-d1.boot-share.n-apps\endcsname{6}
\expandafter\def\csname pct@rev.desc.d3.sspro.p1.C.center-d1.boot-share.n-apps\endcsname{600.0}
\expandafter\def\csname ppt@rev.desc.d3.sspro.p1.C.center-d1.boot-share.n-apps\endcsname{$+$600.0}
\expandafter\def\csname val@rev.desc.d3.sspro.p1.C.center-d1.boot-share.n-boot-valid\endcsname{4000}
\expandafter\def\csname cnt@rev.desc.d3.sspro.p1.C.center-d1.boot-share.n-boot-valid\endcsname{4{,}000}
\expandafter\def\csname pct@rev.desc.d3.sspro.p1.C.center-d1.boot-share.n-boot-valid\endcsname{400000.0}
\expandafter\def\csname ppt@rev.desc.d3.sspro.p1.C.center-d1.boot-share.n-boot-valid\endcsname{$+$400000.0}
\expandafter\def\csname val@rev.desc.d3.sspro.p1.C.center-d1.boot-share.value\endcsname{0.792}
\expandafter\def\csname cnt@rev.desc.d3.sspro.p1.C.center-d1.boot-share.value\endcsname{1}
\expandafter\def\csname pct@rev.desc.d3.sspro.p1.C.center-d1.boot-share.value\endcsname{79.2}
\expandafter\def\csname ppt@rev.desc.d3.sspro.p1.C.center-d1.boot-share.value\endcsname{$+$79.2}
\expandafter\def\csname val@rev.desc.d3.sspro.p1.C.center-d1.n\endcsname{48}
\expandafter\def\csname cnt@rev.desc.d3.sspro.p1.C.center-d1.n\endcsname{48}
\expandafter\def\csname pct@rev.desc.d3.sspro.p1.C.center-d1.n\endcsname{4800.0}
\expandafter\def\csname ppt@rev.desc.d3.sspro.p1.C.center-d1.n\endcsname{$+$4800.0}
\expandafter\def\csname val@rev.desc.d3.sspro.p1.C.center-d1.reproduces-frozen\endcsname{1}
\expandafter\def\csname cnt@rev.desc.d3.sspro.p1.C.center-d1.reproduces-frozen\endcsname{1}
\expandafter\def\csname pct@rev.desc.d3.sspro.p1.C.center-d1.reproduces-frozen\endcsname{100.0}
\expandafter\def\csname ppt@rev.desc.d3.sspro.p1.C.center-d1.reproduces-frozen\endcsname{$+$100.0}
\expandafter\def\csname val@rev.desc.d3.sspro.p1.C.center-d1.resid-agree-stable\endcsname{0.026}
\expandafter\def\csname cnt@rev.desc.d3.sspro.p1.C.center-d1.resid-agree-stable\endcsname{0}
\expandafter\def\csname pct@rev.desc.d3.sspro.p1.C.center-d1.resid-agree-stable\endcsname{2.6}
\expandafter\def\csname ppt@rev.desc.d3.sspro.p1.C.center-d1.resid-agree-stable\endcsname{$+$2.6}
\expandafter\def\csname val@rev.desc.d3.sspro.p1.C.center-d1.resid-disagree\endcsname{0.000}
\expandafter\def\csname cnt@rev.desc.d3.sspro.p1.C.center-d1.resid-disagree\endcsname{0}
\expandafter\def\csname pct@rev.desc.d3.sspro.p1.C.center-d1.resid-disagree\endcsname{0.0}
\expandafter\def\csname ppt@rev.desc.d3.sspro.p1.C.center-d1.resid-disagree\endcsname{0.0}
\expandafter\def\csname val@rev.desc.d3.sspro.p1.C.center-d1.share-agree-stable\endcsname{0.792}
\expandafter\def\csname cnt@rev.desc.d3.sspro.p1.C.center-d1.share-agree-stable\endcsname{1}
\expandafter\def\csname pct@rev.desc.d3.sspro.p1.C.center-d1.share-agree-stable\endcsname{79.2}
\expandafter\def\csname ppt@rev.desc.d3.sspro.p1.C.center-d1.share-agree-stable\endcsname{$+$79.2}
\expandafter\def\csname val@rev.desc.d3.sspro.p1.C.context-d1.N-agree-stable\endcsname{36}
\expandafter\def\csname cnt@rev.desc.d3.sspro.p1.C.context-d1.N-agree-stable\endcsname{36}
\expandafter\def\csname pct@rev.desc.d3.sspro.p1.C.context-d1.N-agree-stable\endcsname{3600.0}
\expandafter\def\csname ppt@rev.desc.d3.sspro.p1.C.context-d1.N-agree-stable\endcsname{$+$3600.0}
\expandafter\def\csname val@rev.desc.d3.sspro.p1.C.context-d1.N-disagree\endcsname{6}
\expandafter\def\csname cnt@rev.desc.d3.sspro.p1.C.context-d1.N-disagree\endcsname{6}
\expandafter\def\csname pct@rev.desc.d3.sspro.p1.C.context-d1.N-disagree\endcsname{600.0}
\expandafter\def\csname ppt@rev.desc.d3.sspro.p1.C.context-d1.N-disagree\endcsname{$+$600.0}
\expandafter\def\csname val@rev.desc.d3.sspro.p1.C.context-d1.boot-resid.ci95.0\endcsname{0.000}
\expandafter\def\csname cnt@rev.desc.d3.sspro.p1.C.context-d1.boot-resid.ci95.0\endcsname{0}
\expandafter\def\csname pct@rev.desc.d3.sspro.p1.C.context-d1.boot-resid.ci95.0\endcsname{0.0}
\expandafter\def\csname ppt@rev.desc.d3.sspro.p1.C.context-d1.boot-resid.ci95.0\endcsname{0.0}
\expandafter\def\csname val@rev.desc.d3.sspro.p1.C.context-d1.boot-resid.ci95.1\endcsname{0.000}
\expandafter\def\csname cnt@rev.desc.d3.sspro.p1.C.context-d1.boot-resid.ci95.1\endcsname{0}
\expandafter\def\csname pct@rev.desc.d3.sspro.p1.C.context-d1.boot-resid.ci95.1\endcsname{0.0}
\expandafter\def\csname ppt@rev.desc.d3.sspro.p1.C.context-d1.boot-resid.ci95.1\endcsname{0.0}
\expandafter\def\csname val@rev.desc.d3.sspro.p1.C.context-d1.boot-resid.n-apps\endcsname{6}
\expandafter\def\csname cnt@rev.desc.d3.sspro.p1.C.context-d1.boot-resid.n-apps\endcsname{6}
\expandafter\def\csname pct@rev.desc.d3.sspro.p1.C.context-d1.boot-resid.n-apps\endcsname{600.0}
\expandafter\def\csname ppt@rev.desc.d3.sspro.p1.C.context-d1.boot-resid.n-apps\endcsname{$+$600.0}
\expandafter\def\csname val@rev.desc.d3.sspro.p1.C.context-d1.boot-resid.n-boot-valid\endcsname{4000}
\expandafter\def\csname cnt@rev.desc.d3.sspro.p1.C.context-d1.boot-resid.n-boot-valid\endcsname{4{,}000}
\expandafter\def\csname pct@rev.desc.d3.sspro.p1.C.context-d1.boot-resid.n-boot-valid\endcsname{400000.0}
\expandafter\def\csname ppt@rev.desc.d3.sspro.p1.C.context-d1.boot-resid.n-boot-valid\endcsname{$+$400000.0}
\expandafter\def\csname val@rev.desc.d3.sspro.p1.C.context-d1.boot-resid.value\endcsname{0.000}
\expandafter\def\csname cnt@rev.desc.d3.sspro.p1.C.context-d1.boot-resid.value\endcsname{0}
\expandafter\def\csname pct@rev.desc.d3.sspro.p1.C.context-d1.boot-resid.value\endcsname{0.0}
\expandafter\def\csname ppt@rev.desc.d3.sspro.p1.C.context-d1.boot-resid.value\endcsname{0.0}
\expandafter\def\csname val@rev.desc.d3.sspro.p1.C.context-d1.boot-share.ci95.0\endcsname{0.583}
\expandafter\def\csname cnt@rev.desc.d3.sspro.p1.C.context-d1.boot-share.ci95.0\endcsname{1}
\expandafter\def\csname pct@rev.desc.d3.sspro.p1.C.context-d1.boot-share.ci95.0\endcsname{58.3}
\expandafter\def\csname ppt@rev.desc.d3.sspro.p1.C.context-d1.boot-share.ci95.0\endcsname{$+$58.3}
\expandafter\def\csname val@rev.desc.d3.sspro.p1.C.context-d1.boot-share.ci95.1\endcsname{0.896}
\expandafter\def\csname cnt@rev.desc.d3.sspro.p1.C.context-d1.boot-share.ci95.1\endcsname{1}
\expandafter\def\csname pct@rev.desc.d3.sspro.p1.C.context-d1.boot-share.ci95.1\endcsname{89.6}
\expandafter\def\csname ppt@rev.desc.d3.sspro.p1.C.context-d1.boot-share.ci95.1\endcsname{$+$89.6}
\expandafter\def\csname val@rev.desc.d3.sspro.p1.C.context-d1.boot-share.n-apps\endcsname{6}
\expandafter\def\csname cnt@rev.desc.d3.sspro.p1.C.context-d1.boot-share.n-apps\endcsname{6}
\expandafter\def\csname pct@rev.desc.d3.sspro.p1.C.context-d1.boot-share.n-apps\endcsname{600.0}
\expandafter\def\csname ppt@rev.desc.d3.sspro.p1.C.context-d1.boot-share.n-apps\endcsname{$+$600.0}
\expandafter\def\csname val@rev.desc.d3.sspro.p1.C.context-d1.boot-share.n-boot-valid\endcsname{4000}
\expandafter\def\csname cnt@rev.desc.d3.sspro.p1.C.context-d1.boot-share.n-boot-valid\endcsname{4{,}000}
\expandafter\def\csname pct@rev.desc.d3.sspro.p1.C.context-d1.boot-share.n-boot-valid\endcsname{400000.0}
\expandafter\def\csname ppt@rev.desc.d3.sspro.p1.C.context-d1.boot-share.n-boot-valid\endcsname{$+$400000.0}
\expandafter\def\csname val@rev.desc.d3.sspro.p1.C.context-d1.boot-share.value\endcsname{0.750}
\expandafter\def\csname cnt@rev.desc.d3.sspro.p1.C.context-d1.boot-share.value\endcsname{1}
\expandafter\def\csname pct@rev.desc.d3.sspro.p1.C.context-d1.boot-share.value\endcsname{75.0}
\expandafter\def\csname ppt@rev.desc.d3.sspro.p1.C.context-d1.boot-share.value\endcsname{$+$75.0}
\expandafter\def\csname val@rev.desc.d3.sspro.p1.C.context-d1.n\endcsname{48}
\expandafter\def\csname cnt@rev.desc.d3.sspro.p1.C.context-d1.n\endcsname{48}
\expandafter\def\csname pct@rev.desc.d3.sspro.p1.C.context-d1.n\endcsname{4800.0}
\expandafter\def\csname ppt@rev.desc.d3.sspro.p1.C.context-d1.n\endcsname{$+$4800.0}
\expandafter\def\csname val@rev.desc.d3.sspro.p1.C.context-d1.reproduces-frozen\endcsname{1}
\expandafter\def\csname cnt@rev.desc.d3.sspro.p1.C.context-d1.reproduces-frozen\endcsname{1}
\expandafter\def\csname pct@rev.desc.d3.sspro.p1.C.context-d1.reproduces-frozen\endcsname{100.0}
\expandafter\def\csname ppt@rev.desc.d3.sspro.p1.C.context-d1.reproduces-frozen\endcsname{$+$100.0}
\expandafter\def\csname val@rev.desc.d3.sspro.p1.C.context-d1.resid-agree-stable\endcsname{0.000}
\expandafter\def\csname cnt@rev.desc.d3.sspro.p1.C.context-d1.resid-agree-stable\endcsname{0}
\expandafter\def\csname pct@rev.desc.d3.sspro.p1.C.context-d1.resid-agree-stable\endcsname{0.0}
\expandafter\def\csname ppt@rev.desc.d3.sspro.p1.C.context-d1.resid-agree-stable\endcsname{0.0}
\expandafter\def\csname val@rev.desc.d3.sspro.p1.C.context-d1.resid-disagree\endcsname{-0.167}
\expandafter\def\csname cnt@rev.desc.d3.sspro.p1.C.context-d1.resid-disagree\endcsname{0}
\expandafter\def\csname pct@rev.desc.d3.sspro.p1.C.context-d1.resid-disagree\endcsname{$-$16.7}
\expandafter\def\csname ppt@rev.desc.d3.sspro.p1.C.context-d1.resid-disagree\endcsname{$-$16.7}
\expandafter\def\csname val@rev.desc.d3.sspro.p1.C.context-d1.share-agree-stable\endcsname{0.750}
\expandafter\def\csname cnt@rev.desc.d3.sspro.p1.C.context-d1.share-agree-stable\endcsname{1}
\expandafter\def\csname pct@rev.desc.d3.sspro.p1.C.context-d1.share-agree-stable\endcsname{75.0}
\expandafter\def\csname ppt@rev.desc.d3.sspro.p1.C.context-d1.share-agree-stable\endcsname{$+$75.0}
\expandafter\def\csname val@rev.desc.d3.sspro.p1.C.dual-d2.N-agree-stable\endcsname{37}
\expandafter\def\csname cnt@rev.desc.d3.sspro.p1.C.dual-d2.N-agree-stable\endcsname{37}
\expandafter\def\csname pct@rev.desc.d3.sspro.p1.C.dual-d2.N-agree-stable\endcsname{3700.0}
\expandafter\def\csname ppt@rev.desc.d3.sspro.p1.C.dual-d2.N-agree-stable\endcsname{$+$3700.0}
\expandafter\def\csname val@rev.desc.d3.sspro.p1.C.dual-d2.N-disagree\endcsname{8}
\expandafter\def\csname cnt@rev.desc.d3.sspro.p1.C.dual-d2.N-disagree\endcsname{8}
\expandafter\def\csname pct@rev.desc.d3.sspro.p1.C.dual-d2.N-disagree\endcsname{800.0}
\expandafter\def\csname ppt@rev.desc.d3.sspro.p1.C.dual-d2.N-disagree\endcsname{$+$800.0}
\expandafter\def\csname val@rev.desc.d3.sspro.p1.C.dual-d2.boot-resid.ci95.0\endcsname{0.000}
\expandafter\def\csname cnt@rev.desc.d3.sspro.p1.C.dual-d2.boot-resid.ci95.0\endcsname{0}
\expandafter\def\csname pct@rev.desc.d3.sspro.p1.C.dual-d2.boot-resid.ci95.0\endcsname{0.0}
\expandafter\def\csname ppt@rev.desc.d3.sspro.p1.C.dual-d2.boot-resid.ci95.0\endcsname{0.0}
\expandafter\def\csname val@rev.desc.d3.sspro.p1.C.dual-d2.boot-resid.ci95.1\endcsname{0.182}
\expandafter\def\csname cnt@rev.desc.d3.sspro.p1.C.dual-d2.boot-resid.ci95.1\endcsname{0}
\expandafter\def\csname pct@rev.desc.d3.sspro.p1.C.dual-d2.boot-resid.ci95.1\endcsname{18.2}
\expandafter\def\csname ppt@rev.desc.d3.sspro.p1.C.dual-d2.boot-resid.ci95.1\endcsname{$+$18.2}
\expandafter\def\csname val@rev.desc.d3.sspro.p1.C.dual-d2.boot-resid.n-apps\endcsname{6}
\expandafter\def\csname cnt@rev.desc.d3.sspro.p1.C.dual-d2.boot-resid.n-apps\endcsname{6}
\expandafter\def\csname pct@rev.desc.d3.sspro.p1.C.dual-d2.boot-resid.n-apps\endcsname{600.0}
\expandafter\def\csname ppt@rev.desc.d3.sspro.p1.C.dual-d2.boot-resid.n-apps\endcsname{$+$600.0}
\expandafter\def\csname val@rev.desc.d3.sspro.p1.C.dual-d2.boot-resid.n-boot-valid\endcsname{4000}
\expandafter\def\csname cnt@rev.desc.d3.sspro.p1.C.dual-d2.boot-resid.n-boot-valid\endcsname{4{,}000}
\expandafter\def\csname pct@rev.desc.d3.sspro.p1.C.dual-d2.boot-resid.n-boot-valid\endcsname{400000.0}
\expandafter\def\csname ppt@rev.desc.d3.sspro.p1.C.dual-d2.boot-resid.n-boot-valid\endcsname{$+$400000.0}
\expandafter\def\csname val@rev.desc.d3.sspro.p1.C.dual-d2.boot-resid.value\endcsname{0.081}
\expandafter\def\csname cnt@rev.desc.d3.sspro.p1.C.dual-d2.boot-resid.value\endcsname{0}
\expandafter\def\csname pct@rev.desc.d3.sspro.p1.C.dual-d2.boot-resid.value\endcsname{8.1}
\expandafter\def\csname ppt@rev.desc.d3.sspro.p1.C.dual-d2.boot-resid.value\endcsname{$+$8.1}
\expandafter\def\csname val@rev.desc.d3.sspro.p1.C.dual-d2.boot-share.ci95.0\endcsname{0.583}
\expandafter\def\csname cnt@rev.desc.d3.sspro.p1.C.dual-d2.boot-share.ci95.0\endcsname{1}
\expandafter\def\csname pct@rev.desc.d3.sspro.p1.C.dual-d2.boot-share.ci95.0\endcsname{58.3}
\expandafter\def\csname ppt@rev.desc.d3.sspro.p1.C.dual-d2.boot-share.ci95.0\endcsname{$+$58.3}
\expandafter\def\csname val@rev.desc.d3.sspro.p1.C.dual-d2.boot-share.ci95.1\endcsname{0.917}
\expandafter\def\csname cnt@rev.desc.d3.sspro.p1.C.dual-d2.boot-share.ci95.1\endcsname{1}
\expandafter\def\csname pct@rev.desc.d3.sspro.p1.C.dual-d2.boot-share.ci95.1\endcsname{91.7}
\expandafter\def\csname ppt@rev.desc.d3.sspro.p1.C.dual-d2.boot-share.ci95.1\endcsname{$+$91.7}
\expandafter\def\csname val@rev.desc.d3.sspro.p1.C.dual-d2.boot-share.n-apps\endcsname{6}
\expandafter\def\csname cnt@rev.desc.d3.sspro.p1.C.dual-d2.boot-share.n-apps\endcsname{6}
\expandafter\def\csname pct@rev.desc.d3.sspro.p1.C.dual-d2.boot-share.n-apps\endcsname{600.0}
\expandafter\def\csname ppt@rev.desc.d3.sspro.p1.C.dual-d2.boot-share.n-apps\endcsname{$+$600.0}
\expandafter\def\csname val@rev.desc.d3.sspro.p1.C.dual-d2.boot-share.n-boot-valid\endcsname{4000}
\expandafter\def\csname cnt@rev.desc.d3.sspro.p1.C.dual-d2.boot-share.n-boot-valid\endcsname{4{,}000}
\expandafter\def\csname pct@rev.desc.d3.sspro.p1.C.dual-d2.boot-share.n-boot-valid\endcsname{400000.0}
\expandafter\def\csname ppt@rev.desc.d3.sspro.p1.C.dual-d2.boot-share.n-boot-valid\endcsname{$+$400000.0}
\expandafter\def\csname val@rev.desc.d3.sspro.p1.C.dual-d2.boot-share.value\endcsname{0.771}
\expandafter\def\csname cnt@rev.desc.d3.sspro.p1.C.dual-d2.boot-share.value\endcsname{1}
\expandafter\def\csname pct@rev.desc.d3.sspro.p1.C.dual-d2.boot-share.value\endcsname{77.1}
\expandafter\def\csname ppt@rev.desc.d3.sspro.p1.C.dual-d2.boot-share.value\endcsname{$+$77.1}
\expandafter\def\csname val@rev.desc.d3.sspro.p1.C.dual-d2.n\endcsname{48}
\expandafter\def\csname cnt@rev.desc.d3.sspro.p1.C.dual-d2.n\endcsname{48}
\expandafter\def\csname pct@rev.desc.d3.sspro.p1.C.dual-d2.n\endcsname{4800.0}
\expandafter\def\csname ppt@rev.desc.d3.sspro.p1.C.dual-d2.n\endcsname{$+$4800.0}
\expandafter\def\csname val@rev.desc.d3.sspro.p1.C.dual-d2.reproduces-frozen\endcsname{1}
\expandafter\def\csname cnt@rev.desc.d3.sspro.p1.C.dual-d2.reproduces-frozen\endcsname{1}
\expandafter\def\csname pct@rev.desc.d3.sspro.p1.C.dual-d2.reproduces-frozen\endcsname{100.0}
\expandafter\def\csname ppt@rev.desc.d3.sspro.p1.C.dual-d2.reproduces-frozen\endcsname{$+$100.0}
\expandafter\def\csname val@rev.desc.d3.sspro.p1.C.dual-d2.resid-agree-stable\endcsname{0.081}
\expandafter\def\csname cnt@rev.desc.d3.sspro.p1.C.dual-d2.resid-agree-stable\endcsname{0}
\expandafter\def\csname pct@rev.desc.d3.sspro.p1.C.dual-d2.resid-agree-stable\endcsname{8.1}
\expandafter\def\csname ppt@rev.desc.d3.sspro.p1.C.dual-d2.resid-agree-stable\endcsname{$+$8.1}
\expandafter\def\csname val@rev.desc.d3.sspro.p1.C.dual-d2.resid-disagree\endcsname{0.000}
\expandafter\def\csname cnt@rev.desc.d3.sspro.p1.C.dual-d2.resid-disagree\endcsname{0}
\expandafter\def\csname pct@rev.desc.d3.sspro.p1.C.dual-d2.resid-disagree\endcsname{0.0}
\expandafter\def\csname ppt@rev.desc.d3.sspro.p1.C.dual-d2.resid-disagree\endcsname{0.0}
\expandafter\def\csname val@rev.desc.d3.sspro.p1.C.dual-d2.share-agree-stable\endcsname{0.771}
\expandafter\def\csname cnt@rev.desc.d3.sspro.p1.C.dual-d2.share-agree-stable\endcsname{1}
\expandafter\def\csname pct@rev.desc.d3.sspro.p1.C.dual-d2.share-agree-stable\endcsname{77.1}
\expandafter\def\csname ppt@rev.desc.d3.sspro.p1.C.dual-d2.share-agree-stable\endcsname{$+$77.1}
\expandafter\def\csname val@rev.desc.d3.sspro.p1.C.grid2-d3.N-agree-stable\endcsname{35}
\expandafter\def\csname cnt@rev.desc.d3.sspro.p1.C.grid2-d3.N-agree-stable\endcsname{35}
\expandafter\def\csname pct@rev.desc.d3.sspro.p1.C.grid2-d3.N-agree-stable\endcsname{3500.0}
\expandafter\def\csname ppt@rev.desc.d3.sspro.p1.C.grid2-d3.N-agree-stable\endcsname{$+$3500.0}
\expandafter\def\csname val@rev.desc.d3.sspro.p1.C.grid2-d3.N-disagree\endcsname{11}
\expandafter\def\csname cnt@rev.desc.d3.sspro.p1.C.grid2-d3.N-disagree\endcsname{11}
\expandafter\def\csname pct@rev.desc.d3.sspro.p1.C.grid2-d3.N-disagree\endcsname{1100.0}
\expandafter\def\csname ppt@rev.desc.d3.sspro.p1.C.grid2-d3.N-disagree\endcsname{$+$1100.0}
\expandafter\def\csname val@rev.desc.d3.sspro.p1.C.grid2-d3.boot-resid.ci95.0\endcsname{-0.118}
\expandafter\def\csname cnt@rev.desc.d3.sspro.p1.C.grid2-d3.boot-resid.ci95.0\endcsname{0}
\expandafter\def\csname pct@rev.desc.d3.sspro.p1.C.grid2-d3.boot-resid.ci95.0\endcsname{$-$11.8}
\expandafter\def\csname ppt@rev.desc.d3.sspro.p1.C.grid2-d3.boot-resid.ci95.0\endcsname{$-$11.8}
\expandafter\def\csname val@rev.desc.d3.sspro.p1.C.grid2-d3.boot-resid.ci95.1\endcsname{0.000}
\expandafter\def\csname cnt@rev.desc.d3.sspro.p1.C.grid2-d3.boot-resid.ci95.1\endcsname{0}
\expandafter\def\csname pct@rev.desc.d3.sspro.p1.C.grid2-d3.boot-resid.ci95.1\endcsname{0.0}
\expandafter\def\csname ppt@rev.desc.d3.sspro.p1.C.grid2-d3.boot-resid.ci95.1\endcsname{0.0}
\expandafter\def\csname val@rev.desc.d3.sspro.p1.C.grid2-d3.boot-resid.n-apps\endcsname{6}
\expandafter\def\csname cnt@rev.desc.d3.sspro.p1.C.grid2-d3.boot-resid.n-apps\endcsname{6}
\expandafter\def\csname pct@rev.desc.d3.sspro.p1.C.grid2-d3.boot-resid.n-apps\endcsname{600.0}
\expandafter\def\csname ppt@rev.desc.d3.sspro.p1.C.grid2-d3.boot-resid.n-apps\endcsname{$+$600.0}
\expandafter\def\csname val@rev.desc.d3.sspro.p1.C.grid2-d3.boot-resid.n-boot-valid\endcsname{4000}
\expandafter\def\csname cnt@rev.desc.d3.sspro.p1.C.grid2-d3.boot-resid.n-boot-valid\endcsname{4{,}000}
\expandafter\def\csname pct@rev.desc.d3.sspro.p1.C.grid2-d3.boot-resid.n-boot-valid\endcsname{400000.0}
\expandafter\def\csname ppt@rev.desc.d3.sspro.p1.C.grid2-d3.boot-resid.n-boot-valid\endcsname{$+$400000.0}
\expandafter\def\csname val@rev.desc.d3.sspro.p1.C.grid2-d3.boot-resid.value\endcsname{-0.057}
\expandafter\def\csname cnt@rev.desc.d3.sspro.p1.C.grid2-d3.boot-resid.value\endcsname{0}
\expandafter\def\csname pct@rev.desc.d3.sspro.p1.C.grid2-d3.boot-resid.value\endcsname{$-$5.7}
\expandafter\def\csname ppt@rev.desc.d3.sspro.p1.C.grid2-d3.boot-resid.value\endcsname{$-$5.7}
\expandafter\def\csname val@rev.desc.d3.sspro.p1.C.grid2-d3.boot-share.ci95.0\endcsname{0.583}
\expandafter\def\csname cnt@rev.desc.d3.sspro.p1.C.grid2-d3.boot-share.ci95.0\endcsname{1}
\expandafter\def\csname pct@rev.desc.d3.sspro.p1.C.grid2-d3.boot-share.ci95.0\endcsname{58.3}
\expandafter\def\csname ppt@rev.desc.d3.sspro.p1.C.grid2-d3.boot-share.ci95.0\endcsname{$+$58.3}
\expandafter\def\csname val@rev.desc.d3.sspro.p1.C.grid2-d3.boot-share.ci95.1\endcsname{0.833}
\expandafter\def\csname cnt@rev.desc.d3.sspro.p1.C.grid2-d3.boot-share.ci95.1\endcsname{1}
\expandafter\def\csname pct@rev.desc.d3.sspro.p1.C.grid2-d3.boot-share.ci95.1\endcsname{83.3}
\expandafter\def\csname ppt@rev.desc.d3.sspro.p1.C.grid2-d3.boot-share.ci95.1\endcsname{$+$83.3}
\expandafter\def\csname val@rev.desc.d3.sspro.p1.C.grid2-d3.boot-share.n-apps\endcsname{6}
\expandafter\def\csname cnt@rev.desc.d3.sspro.p1.C.grid2-d3.boot-share.n-apps\endcsname{6}
\expandafter\def\csname pct@rev.desc.d3.sspro.p1.C.grid2-d3.boot-share.n-apps\endcsname{600.0}
\expandafter\def\csname ppt@rev.desc.d3.sspro.p1.C.grid2-d3.boot-share.n-apps\endcsname{$+$600.0}
\expandafter\def\csname val@rev.desc.d3.sspro.p1.C.grid2-d3.boot-share.n-boot-valid\endcsname{4000}
\expandafter\def\csname cnt@rev.desc.d3.sspro.p1.C.grid2-d3.boot-share.n-boot-valid\endcsname{4{,}000}
\expandafter\def\csname pct@rev.desc.d3.sspro.p1.C.grid2-d3.boot-share.n-boot-valid\endcsname{400000.0}
\expandafter\def\csname ppt@rev.desc.d3.sspro.p1.C.grid2-d3.boot-share.n-boot-valid\endcsname{$+$400000.0}
\expandafter\def\csname val@rev.desc.d3.sspro.p1.C.grid2-d3.boot-share.value\endcsname{0.729}
\expandafter\def\csname cnt@rev.desc.d3.sspro.p1.C.grid2-d3.boot-share.value\endcsname{1}
\expandafter\def\csname pct@rev.desc.d3.sspro.p1.C.grid2-d3.boot-share.value\endcsname{72.9}
\expandafter\def\csname ppt@rev.desc.d3.sspro.p1.C.grid2-d3.boot-share.value\endcsname{$+$72.9}
\expandafter\def\csname val@rev.desc.d3.sspro.p1.C.grid2-d3.n\endcsname{48}
\expandafter\def\csname cnt@rev.desc.d3.sspro.p1.C.grid2-d3.n\endcsname{48}
\expandafter\def\csname pct@rev.desc.d3.sspro.p1.C.grid2-d3.n\endcsname{4800.0}
\expandafter\def\csname ppt@rev.desc.d3.sspro.p1.C.grid2-d3.n\endcsname{$+$4800.0}
\expandafter\def\csname val@rev.desc.d3.sspro.p1.C.grid2-d3.reproduces-frozen\endcsname{1}
\expandafter\def\csname cnt@rev.desc.d3.sspro.p1.C.grid2-d3.reproduces-frozen\endcsname{1}
\expandafter\def\csname pct@rev.desc.d3.sspro.p1.C.grid2-d3.reproduces-frozen\endcsname{100.0}
\expandafter\def\csname ppt@rev.desc.d3.sspro.p1.C.grid2-d3.reproduces-frozen\endcsname{$+$100.0}
\expandafter\def\csname val@rev.desc.d3.sspro.p1.C.grid2-d3.resid-agree-stable\endcsname{-0.057}
\expandafter\def\csname cnt@rev.desc.d3.sspro.p1.C.grid2-d3.resid-agree-stable\endcsname{0}
\expandafter\def\csname pct@rev.desc.d3.sspro.p1.C.grid2-d3.resid-agree-stable\endcsname{$-$5.7}
\expandafter\def\csname ppt@rev.desc.d3.sspro.p1.C.grid2-d3.resid-agree-stable\endcsname{$-$5.7}
\expandafter\def\csname val@rev.desc.d3.sspro.p1.C.grid2-d3.resid-disagree\endcsname{0.182}
\expandafter\def\csname cnt@rev.desc.d3.sspro.p1.C.grid2-d3.resid-disagree\endcsname{0}
\expandafter\def\csname pct@rev.desc.d3.sspro.p1.C.grid2-d3.resid-disagree\endcsname{18.2}
\expandafter\def\csname ppt@rev.desc.d3.sspro.p1.C.grid2-d3.resid-disagree\endcsname{$+$18.2}
\expandafter\def\csname val@rev.desc.d3.sspro.p1.C.grid2-d3.share-agree-stable\endcsname{0.729}
\expandafter\def\csname cnt@rev.desc.d3.sspro.p1.C.grid2-d3.share-agree-stable\endcsname{1}
\expandafter\def\csname pct@rev.desc.d3.sspro.p1.C.grid2-d3.share-agree-stable\endcsname{72.9}
\expandafter\def\csname ppt@rev.desc.d3.sspro.p1.C.grid2-d3.share-agree-stable\endcsname{$+$72.9}
\expandafter\def\csname val@rev.desc.d3.sspro.p1.C.zoomclick-d3.N-agree-stable\endcsname{38}
\expandafter\def\csname cnt@rev.desc.d3.sspro.p1.C.zoomclick-d3.N-agree-stable\endcsname{38}
\expandafter\def\csname pct@rev.desc.d3.sspro.p1.C.zoomclick-d3.N-agree-stable\endcsname{3800.0}
\expandafter\def\csname ppt@rev.desc.d3.sspro.p1.C.zoomclick-d3.N-agree-stable\endcsname{$+$3800.0}
\expandafter\def\csname val@rev.desc.d3.sspro.p1.C.zoomclick-d3.N-disagree\endcsname{7}
\expandafter\def\csname cnt@rev.desc.d3.sspro.p1.C.zoomclick-d3.N-disagree\endcsname{7}
\expandafter\def\csname pct@rev.desc.d3.sspro.p1.C.zoomclick-d3.N-disagree\endcsname{700.0}
\expandafter\def\csname ppt@rev.desc.d3.sspro.p1.C.zoomclick-d3.N-disagree\endcsname{$+$700.0}
\expandafter\def\csname val@rev.desc.d3.sspro.p1.C.zoomclick-d3.boot-resid.ci95.0\endcsname{0.000}
\expandafter\def\csname cnt@rev.desc.d3.sspro.p1.C.zoomclick-d3.boot-resid.ci95.0\endcsname{0}
\expandafter\def\csname pct@rev.desc.d3.sspro.p1.C.zoomclick-d3.boot-resid.ci95.0\endcsname{0.0}
\expandafter\def\csname ppt@rev.desc.d3.sspro.p1.C.zoomclick-d3.boot-resid.ci95.0\endcsname{0.0}
\expandafter\def\csname val@rev.desc.d3.sspro.p1.C.zoomclick-d3.boot-resid.ci95.1\endcsname{0.000}
\expandafter\def\csname cnt@rev.desc.d3.sspro.p1.C.zoomclick-d3.boot-resid.ci95.1\endcsname{0}
\expandafter\def\csname pct@rev.desc.d3.sspro.p1.C.zoomclick-d3.boot-resid.ci95.1\endcsname{0.0}
\expandafter\def\csname ppt@rev.desc.d3.sspro.p1.C.zoomclick-d3.boot-resid.ci95.1\endcsname{0.0}
\expandafter\def\csname val@rev.desc.d3.sspro.p1.C.zoomclick-d3.boot-resid.n-apps\endcsname{6}
\expandafter\def\csname cnt@rev.desc.d3.sspro.p1.C.zoomclick-d3.boot-resid.n-apps\endcsname{6}
\expandafter\def\csname pct@rev.desc.d3.sspro.p1.C.zoomclick-d3.boot-resid.n-apps\endcsname{600.0}
\expandafter\def\csname ppt@rev.desc.d3.sspro.p1.C.zoomclick-d3.boot-resid.n-apps\endcsname{$+$600.0}
\expandafter\def\csname val@rev.desc.d3.sspro.p1.C.zoomclick-d3.boot-resid.n-boot-valid\endcsname{4000}
\expandafter\def\csname cnt@rev.desc.d3.sspro.p1.C.zoomclick-d3.boot-resid.n-boot-valid\endcsname{4{,}000}
\expandafter\def\csname pct@rev.desc.d3.sspro.p1.C.zoomclick-d3.boot-resid.n-boot-valid\endcsname{400000.0}
\expandafter\def\csname ppt@rev.desc.d3.sspro.p1.C.zoomclick-d3.boot-resid.n-boot-valid\endcsname{$+$400000.0}
\expandafter\def\csname val@rev.desc.d3.sspro.p1.C.zoomclick-d3.boot-resid.value\endcsname{0.000}
\expandafter\def\csname cnt@rev.desc.d3.sspro.p1.C.zoomclick-d3.boot-resid.value\endcsname{0}
\expandafter\def\csname pct@rev.desc.d3.sspro.p1.C.zoomclick-d3.boot-resid.value\endcsname{0.0}
\expandafter\def\csname ppt@rev.desc.d3.sspro.p1.C.zoomclick-d3.boot-resid.value\endcsname{0.0}
\expandafter\def\csname val@rev.desc.d3.sspro.p1.C.zoomclick-d3.boot-share.ci95.0\endcsname{0.625}
\expandafter\def\csname cnt@rev.desc.d3.sspro.p1.C.zoomclick-d3.boot-share.ci95.0\endcsname{1}
\expandafter\def\csname pct@rev.desc.d3.sspro.p1.C.zoomclick-d3.boot-share.ci95.0\endcsname{62.5}
\expandafter\def\csname ppt@rev.desc.d3.sspro.p1.C.zoomclick-d3.boot-share.ci95.0\endcsname{$+$62.5}
\expandafter\def\csname val@rev.desc.d3.sspro.p1.C.zoomclick-d3.boot-share.ci95.1\endcsname{0.917}
\expandafter\def\csname cnt@rev.desc.d3.sspro.p1.C.zoomclick-d3.boot-share.ci95.1\endcsname{1}
\expandafter\def\csname pct@rev.desc.d3.sspro.p1.C.zoomclick-d3.boot-share.ci95.1\endcsname{91.7}
\expandafter\def\csname ppt@rev.desc.d3.sspro.p1.C.zoomclick-d3.boot-share.ci95.1\endcsname{$+$91.7}
\expandafter\def\csname val@rev.desc.d3.sspro.p1.C.zoomclick-d3.boot-share.n-apps\endcsname{6}
\expandafter\def\csname cnt@rev.desc.d3.sspro.p1.C.zoomclick-d3.boot-share.n-apps\endcsname{6}
\expandafter\def\csname pct@rev.desc.d3.sspro.p1.C.zoomclick-d3.boot-share.n-apps\endcsname{600.0}
\expandafter\def\csname ppt@rev.desc.d3.sspro.p1.C.zoomclick-d3.boot-share.n-apps\endcsname{$+$600.0}
\expandafter\def\csname val@rev.desc.d3.sspro.p1.C.zoomclick-d3.boot-share.n-boot-valid\endcsname{4000}
\expandafter\def\csname cnt@rev.desc.d3.sspro.p1.C.zoomclick-d3.boot-share.n-boot-valid\endcsname{4{,}000}
\expandafter\def\csname pct@rev.desc.d3.sspro.p1.C.zoomclick-d3.boot-share.n-boot-valid\endcsname{400000.0}
\expandafter\def\csname ppt@rev.desc.d3.sspro.p1.C.zoomclick-d3.boot-share.n-boot-valid\endcsname{$+$400000.0}
\expandafter\def\csname val@rev.desc.d3.sspro.p1.C.zoomclick-d3.boot-share.value\endcsname{0.792}
\expandafter\def\csname cnt@rev.desc.d3.sspro.p1.C.zoomclick-d3.boot-share.value\endcsname{1}
\expandafter\def\csname pct@rev.desc.d3.sspro.p1.C.zoomclick-d3.boot-share.value\endcsname{79.2}
\expandafter\def\csname ppt@rev.desc.d3.sspro.p1.C.zoomclick-d3.boot-share.value\endcsname{$+$79.2}
\expandafter\def\csname val@rev.desc.d3.sspro.p1.C.zoomclick-d3.n\endcsname{48}
\expandafter\def\csname cnt@rev.desc.d3.sspro.p1.C.zoomclick-d3.n\endcsname{48}
\expandafter\def\csname pct@rev.desc.d3.sspro.p1.C.zoomclick-d3.n\endcsname{4800.0}
\expandafter\def\csname ppt@rev.desc.d3.sspro.p1.C.zoomclick-d3.n\endcsname{$+$4800.0}
\expandafter\def\csname val@rev.desc.d3.sspro.p1.C.zoomclick-d3.reproduces-frozen\endcsname{1}
\expandafter\def\csname cnt@rev.desc.d3.sspro.p1.C.zoomclick-d3.reproduces-frozen\endcsname{1}
\expandafter\def\csname pct@rev.desc.d3.sspro.p1.C.zoomclick-d3.reproduces-frozen\endcsname{100.0}
\expandafter\def\csname ppt@rev.desc.d3.sspro.p1.C.zoomclick-d3.reproduces-frozen\endcsname{$+$100.0}
\expandafter\def\csname val@rev.desc.d3.sspro.p1.C.zoomclick-d3.resid-agree-stable\endcsname{0.000}
\expandafter\def\csname cnt@rev.desc.d3.sspro.p1.C.zoomclick-d3.resid-agree-stable\endcsname{0}
\expandafter\def\csname pct@rev.desc.d3.sspro.p1.C.zoomclick-d3.resid-agree-stable\endcsname{0.0}
\expandafter\def\csname ppt@rev.desc.d3.sspro.p1.C.zoomclick-d3.resid-agree-stable\endcsname{0.0}
\expandafter\def\csname val@rev.desc.d3.sspro.p1.C.zoomclick-d3.resid-disagree\endcsname{0.000}
\expandafter\def\csname cnt@rev.desc.d3.sspro.p1.C.zoomclick-d3.resid-disagree\endcsname{0}
\expandafter\def\csname pct@rev.desc.d3.sspro.p1.C.zoomclick-d3.resid-disagree\endcsname{0.0}
\expandafter\def\csname ppt@rev.desc.d3.sspro.p1.C.zoomclick-d3.resid-disagree\endcsname{0.0}
\expandafter\def\csname val@rev.desc.d3.sspro.p1.C.zoomclick-d3.share-agree-stable\endcsname{0.792}
\expandafter\def\csname cnt@rev.desc.d3.sspro.p1.C.zoomclick-d3.share-agree-stable\endcsname{1}
\expandafter\def\csname pct@rev.desc.d3.sspro.p1.C.zoomclick-d3.share-agree-stable\endcsname{79.2}
\expandafter\def\csname ppt@rev.desc.d3.sspro.p1.C.zoomclick-d3.share-agree-stable\endcsname{$+$79.2}
\expandafter\def\csname val@rev.desc.d3.sspro.p2.C.catalog-pooled.N-agree-stable\endcsname{160}
\expandafter\def\csname cnt@rev.desc.d3.sspro.p2.C.catalog-pooled.N-agree-stable\endcsname{160}
\expandafter\def\csname pct@rev.desc.d3.sspro.p2.C.catalog-pooled.N-agree-stable\endcsname{16000.0}
\expandafter\def\csname ppt@rev.desc.d3.sspro.p2.C.catalog-pooled.N-agree-stable\endcsname{$+$16000.0}
\expandafter\def\csname val@rev.desc.d3.sspro.p2.C.catalog-pooled.N-disagree\endcsname{28}
\expandafter\def\csname cnt@rev.desc.d3.sspro.p2.C.catalog-pooled.N-disagree\endcsname{28}
\expandafter\def\csname pct@rev.desc.d3.sspro.p2.C.catalog-pooled.N-disagree\endcsname{2800.0}
\expandafter\def\csname ppt@rev.desc.d3.sspro.p2.C.catalog-pooled.N-disagree\endcsname{$+$2800.0}
\expandafter\def\csname val@rev.desc.d3.sspro.p2.C.catalog-pooled.boot-resid.ci95.0\endcsname{-0.043}
\expandafter\def\csname cnt@rev.desc.d3.sspro.p2.C.catalog-pooled.boot-resid.ci95.0\endcsname{0}
\expandafter\def\csname pct@rev.desc.d3.sspro.p2.C.catalog-pooled.boot-resid.ci95.0\endcsname{$-$4.3}
\expandafter\def\csname ppt@rev.desc.d3.sspro.p2.C.catalog-pooled.boot-resid.ci95.0\endcsname{$-$4.3}
\expandafter\def\csname val@rev.desc.d3.sspro.p2.C.catalog-pooled.boot-resid.ci95.1\endcsname{0.013}
\expandafter\def\csname cnt@rev.desc.d3.sspro.p2.C.catalog-pooled.boot-resid.ci95.1\endcsname{0}
\expandafter\def\csname pct@rev.desc.d3.sspro.p2.C.catalog-pooled.boot-resid.ci95.1\endcsname{1.3}
\expandafter\def\csname ppt@rev.desc.d3.sspro.p2.C.catalog-pooled.boot-resid.ci95.1\endcsname{$+$1.3}
\expandafter\def\csname val@rev.desc.d3.sspro.p2.C.catalog-pooled.boot-resid.n-apps\endcsname{6}
\expandafter\def\csname cnt@rev.desc.d3.sspro.p2.C.catalog-pooled.boot-resid.n-apps\endcsname{6}
\expandafter\def\csname pct@rev.desc.d3.sspro.p2.C.catalog-pooled.boot-resid.n-apps\endcsname{600.0}
\expandafter\def\csname ppt@rev.desc.d3.sspro.p2.C.catalog-pooled.boot-resid.n-apps\endcsname{$+$600.0}
\expandafter\def\csname val@rev.desc.d3.sspro.p2.C.catalog-pooled.boot-resid.n-boot-valid\endcsname{4000}
\expandafter\def\csname cnt@rev.desc.d3.sspro.p2.C.catalog-pooled.boot-resid.n-boot-valid\endcsname{4{,}000}
\expandafter\def\csname pct@rev.desc.d3.sspro.p2.C.catalog-pooled.boot-resid.n-boot-valid\endcsname{400000.0}
\expandafter\def\csname ppt@rev.desc.d3.sspro.p2.C.catalog-pooled.boot-resid.n-boot-valid\endcsname{$+$400000.0}
\expandafter\def\csname val@rev.desc.d3.sspro.p2.C.catalog-pooled.boot-resid.value\endcsname{-0.013}
\expandafter\def\csname cnt@rev.desc.d3.sspro.p2.C.catalog-pooled.boot-resid.value\endcsname{0}
\expandafter\def\csname pct@rev.desc.d3.sspro.p2.C.catalog-pooled.boot-resid.value\endcsname{$-$1.2}
\expandafter\def\csname ppt@rev.desc.d3.sspro.p2.C.catalog-pooled.boot-resid.value\endcsname{$-$1.2}
\expandafter\def\csname val@rev.desc.d3.sspro.p2.C.catalog-pooled.boot-share.ci95.0\endcsname{0.609}
\expandafter\def\csname cnt@rev.desc.d3.sspro.p2.C.catalog-pooled.boot-share.ci95.0\endcsname{1}
\expandafter\def\csname pct@rev.desc.d3.sspro.p2.C.catalog-pooled.boot-share.ci95.0\endcsname{60.9}
\expandafter\def\csname ppt@rev.desc.d3.sspro.p2.C.catalog-pooled.boot-share.ci95.0\endcsname{$+$60.9}
\expandafter\def\csname val@rev.desc.d3.sspro.p2.C.catalog-pooled.boot-share.ci95.1\endcsname{0.964}
\expandafter\def\csname cnt@rev.desc.d3.sspro.p2.C.catalog-pooled.boot-share.ci95.1\endcsname{1}
\expandafter\def\csname pct@rev.desc.d3.sspro.p2.C.catalog-pooled.boot-share.ci95.1\endcsname{96.4}
\expandafter\def\csname ppt@rev.desc.d3.sspro.p2.C.catalog-pooled.boot-share.ci95.1\endcsname{$+$96.4}
\expandafter\def\csname val@rev.desc.d3.sspro.p2.C.catalog-pooled.boot-share.n-apps\endcsname{6}
\expandafter\def\csname cnt@rev.desc.d3.sspro.p2.C.catalog-pooled.boot-share.n-apps\endcsname{6}
\expandafter\def\csname pct@rev.desc.d3.sspro.p2.C.catalog-pooled.boot-share.n-apps\endcsname{600.0}
\expandafter\def\csname ppt@rev.desc.d3.sspro.p2.C.catalog-pooled.boot-share.n-apps\endcsname{$+$600.0}
\expandafter\def\csname val@rev.desc.d3.sspro.p2.C.catalog-pooled.boot-share.n-boot-valid\endcsname{4000}
\expandafter\def\csname cnt@rev.desc.d3.sspro.p2.C.catalog-pooled.boot-share.n-boot-valid\endcsname{4{,}000}
\expandafter\def\csname pct@rev.desc.d3.sspro.p2.C.catalog-pooled.boot-share.n-boot-valid\endcsname{400000.0}
\expandafter\def\csname ppt@rev.desc.d3.sspro.p2.C.catalog-pooled.boot-share.n-boot-valid\endcsname{$+$400000.0}
\expandafter\def\csname val@rev.desc.d3.sspro.p2.C.catalog-pooled.boot-share.value\endcsname{0.833}
\expandafter\def\csname cnt@rev.desc.d3.sspro.p2.C.catalog-pooled.boot-share.value\endcsname{1}
\expandafter\def\csname pct@rev.desc.d3.sspro.p2.C.catalog-pooled.boot-share.value\endcsname{83.3}
\expandafter\def\csname ppt@rev.desc.d3.sspro.p2.C.catalog-pooled.boot-share.value\endcsname{$+$83.3}
\expandafter\def\csname val@rev.desc.d3.sspro.p2.C.catalog-pooled.gap\endcsname{0.120}
\expandafter\def\csname cnt@rev.desc.d3.sspro.p2.C.catalog-pooled.gap\endcsname{0}
\expandafter\def\csname pct@rev.desc.d3.sspro.p2.C.catalog-pooled.gap\endcsname{12.0}
\expandafter\def\csname ppt@rev.desc.d3.sspro.p2.C.catalog-pooled.gap\endcsname{$+$12.0}
\expandafter\def\csname val@rev.desc.d3.sspro.p2.C.catalog-pooled.n\endcsname{192}
\expandafter\def\csname cnt@rev.desc.d3.sspro.p2.C.catalog-pooled.n\endcsname{192}
\expandafter\def\csname pct@rev.desc.d3.sspro.p2.C.catalog-pooled.n\endcsname{19200.0}
\expandafter\def\csname ppt@rev.desc.d3.sspro.p2.C.catalog-pooled.n\endcsname{$+$19200.0}
\expandafter\def\csname val@rev.desc.d3.sspro.p2.C.catalog-pooled.reproduces-frozen\endcsname{1}
\expandafter\def\csname cnt@rev.desc.d3.sspro.p2.C.catalog-pooled.reproduces-frozen\endcsname{1}
\expandafter\def\csname pct@rev.desc.d3.sspro.p2.C.catalog-pooled.reproduces-frozen\endcsname{100.0}
\expandafter\def\csname ppt@rev.desc.d3.sspro.p2.C.catalog-pooled.reproduces-frozen\endcsname{$+$100.0}
\expandafter\def\csname val@rev.desc.d3.sspro.p2.C.catalog-pooled.resid-agree-stable\endcsname{-0.013}
\expandafter\def\csname cnt@rev.desc.d3.sspro.p2.C.catalog-pooled.resid-agree-stable\endcsname{0}
\expandafter\def\csname pct@rev.desc.d3.sspro.p2.C.catalog-pooled.resid-agree-stable\endcsname{$-$1.2}
\expandafter\def\csname ppt@rev.desc.d3.sspro.p2.C.catalog-pooled.resid-agree-stable\endcsname{$-$1.2}
\expandafter\def\csname val@rev.desc.d3.sspro.p2.C.catalog-pooled.resid-disagree\endcsname{0.107}
\expandafter\def\csname cnt@rev.desc.d3.sspro.p2.C.catalog-pooled.resid-disagree\endcsname{0}
\expandafter\def\csname pct@rev.desc.d3.sspro.p2.C.catalog-pooled.resid-disagree\endcsname{10.7}
\expandafter\def\csname ppt@rev.desc.d3.sspro.p2.C.catalog-pooled.resid-disagree\endcsname{$+$10.7}
\expandafter\def\csname val@rev.desc.d3.sspro.p2.C.catalog-pooled.share-agree-stable\endcsname{0.833}
\expandafter\def\csname cnt@rev.desc.d3.sspro.p2.C.catalog-pooled.share-agree-stable\endcsname{1}
\expandafter\def\csname pct@rev.desc.d3.sspro.p2.C.catalog-pooled.share-agree-stable\endcsname{83.3}
\expandafter\def\csname ppt@rev.desc.d3.sspro.p2.C.catalog-pooled.share-agree-stable\endcsname{$+$83.3}
\expandafter\def\csname val@rev.desc.d3.sspro.p2.C.center-d1.N-agree-stable\endcsname{42}
\expandafter\def\csname cnt@rev.desc.d3.sspro.p2.C.center-d1.N-agree-stable\endcsname{42}
\expandafter\def\csname pct@rev.desc.d3.sspro.p2.C.center-d1.N-agree-stable\endcsname{4200.0}
\expandafter\def\csname ppt@rev.desc.d3.sspro.p2.C.center-d1.N-agree-stable\endcsname{$+$4200.0}
\expandafter\def\csname val@rev.desc.d3.sspro.p2.C.center-d1.N-disagree\endcsname{6}
\expandafter\def\csname cnt@rev.desc.d3.sspro.p2.C.center-d1.N-disagree\endcsname{6}
\expandafter\def\csname pct@rev.desc.d3.sspro.p2.C.center-d1.N-disagree\endcsname{600.0}
\expandafter\def\csname ppt@rev.desc.d3.sspro.p2.C.center-d1.N-disagree\endcsname{$+$600.0}
\expandafter\def\csname val@rev.desc.d3.sspro.p2.C.center-d1.boot-resid.ci95.0\endcsname{0.000}
\expandafter\def\csname cnt@rev.desc.d3.sspro.p2.C.center-d1.boot-resid.ci95.0\endcsname{0}
\expandafter\def\csname pct@rev.desc.d3.sspro.p2.C.center-d1.boot-resid.ci95.0\endcsname{0.0}
\expandafter\def\csname ppt@rev.desc.d3.sspro.p2.C.center-d1.boot-resid.ci95.0\endcsname{0.0}
\expandafter\def\csname val@rev.desc.d3.sspro.p2.C.center-d1.boot-resid.ci95.1\endcsname{0.071}
\expandafter\def\csname cnt@rev.desc.d3.sspro.p2.C.center-d1.boot-resid.ci95.1\endcsname{0}
\expandafter\def\csname pct@rev.desc.d3.sspro.p2.C.center-d1.boot-resid.ci95.1\endcsname{7.1}
\expandafter\def\csname ppt@rev.desc.d3.sspro.p2.C.center-d1.boot-resid.ci95.1\endcsname{$+$7.1}
\expandafter\def\csname val@rev.desc.d3.sspro.p2.C.center-d1.boot-resid.n-apps\endcsname{6}
\expandafter\def\csname cnt@rev.desc.d3.sspro.p2.C.center-d1.boot-resid.n-apps\endcsname{6}
\expandafter\def\csname pct@rev.desc.d3.sspro.p2.C.center-d1.boot-resid.n-apps\endcsname{600.0}
\expandafter\def\csname ppt@rev.desc.d3.sspro.p2.C.center-d1.boot-resid.n-apps\endcsname{$+$600.0}
\expandafter\def\csname val@rev.desc.d3.sspro.p2.C.center-d1.boot-resid.n-boot-valid\endcsname{4000}
\expandafter\def\csname cnt@rev.desc.d3.sspro.p2.C.center-d1.boot-resid.n-boot-valid\endcsname{4{,}000}
\expandafter\def\csname pct@rev.desc.d3.sspro.p2.C.center-d1.boot-resid.n-boot-valid\endcsname{400000.0}
\expandafter\def\csname ppt@rev.desc.d3.sspro.p2.C.center-d1.boot-resid.n-boot-valid\endcsname{$+$400000.0}
\expandafter\def\csname val@rev.desc.d3.sspro.p2.C.center-d1.boot-resid.value\endcsname{0.024}
\expandafter\def\csname cnt@rev.desc.d3.sspro.p2.C.center-d1.boot-resid.value\endcsname{0}
\expandafter\def\csname pct@rev.desc.d3.sspro.p2.C.center-d1.boot-resid.value\endcsname{2.4}
\expandafter\def\csname ppt@rev.desc.d3.sspro.p2.C.center-d1.boot-resid.value\endcsname{$+$2.4}
\expandafter\def\csname val@rev.desc.d3.sspro.p2.C.center-d1.boot-share.ci95.0\endcsname{0.667}
\expandafter\def\csname cnt@rev.desc.d3.sspro.p2.C.center-d1.boot-share.ci95.0\endcsname{1}
\expandafter\def\csname pct@rev.desc.d3.sspro.p2.C.center-d1.boot-share.ci95.0\endcsname{66.7}
\expandafter\def\csname ppt@rev.desc.d3.sspro.p2.C.center-d1.boot-share.ci95.0\endcsname{$+$66.7}
\expandafter\def\csname val@rev.desc.d3.sspro.p2.C.center-d1.boot-share.ci95.1\endcsname{1.000}
\expandafter\def\csname cnt@rev.desc.d3.sspro.p2.C.center-d1.boot-share.ci95.1\endcsname{1}
\expandafter\def\csname pct@rev.desc.d3.sspro.p2.C.center-d1.boot-share.ci95.1\endcsname{100.0}
\expandafter\def\csname ppt@rev.desc.d3.sspro.p2.C.center-d1.boot-share.ci95.1\endcsname{$+$100.0}
\expandafter\def\csname val@rev.desc.d3.sspro.p2.C.center-d1.boot-share.n-apps\endcsname{6}
\expandafter\def\csname cnt@rev.desc.d3.sspro.p2.C.center-d1.boot-share.n-apps\endcsname{6}
\expandafter\def\csname pct@rev.desc.d3.sspro.p2.C.center-d1.boot-share.n-apps\endcsname{600.0}
\expandafter\def\csname ppt@rev.desc.d3.sspro.p2.C.center-d1.boot-share.n-apps\endcsname{$+$600.0}
\expandafter\def\csname val@rev.desc.d3.sspro.p2.C.center-d1.boot-share.n-boot-valid\endcsname{4000}
\expandafter\def\csname cnt@rev.desc.d3.sspro.p2.C.center-d1.boot-share.n-boot-valid\endcsname{4{,}000}
\expandafter\def\csname pct@rev.desc.d3.sspro.p2.C.center-d1.boot-share.n-boot-valid\endcsname{400000.0}
\expandafter\def\csname ppt@rev.desc.d3.sspro.p2.C.center-d1.boot-share.n-boot-valid\endcsname{$+$400000.0}
\expandafter\def\csname val@rev.desc.d3.sspro.p2.C.center-d1.boot-share.value\endcsname{0.875}
\expandafter\def\csname cnt@rev.desc.d3.sspro.p2.C.center-d1.boot-share.value\endcsname{1}
\expandafter\def\csname pct@rev.desc.d3.sspro.p2.C.center-d1.boot-share.value\endcsname{87.5}
\expandafter\def\csname ppt@rev.desc.d3.sspro.p2.C.center-d1.boot-share.value\endcsname{$+$87.5}
\expandafter\def\csname val@rev.desc.d3.sspro.p2.C.center-d1.n\endcsname{48}
\expandafter\def\csname cnt@rev.desc.d3.sspro.p2.C.center-d1.n\endcsname{48}
\expandafter\def\csname pct@rev.desc.d3.sspro.p2.C.center-d1.n\endcsname{4800.0}
\expandafter\def\csname ppt@rev.desc.d3.sspro.p2.C.center-d1.n\endcsname{$+$4800.0}
\expandafter\def\csname val@rev.desc.d3.sspro.p2.C.center-d1.reproduces-frozen\endcsname{1}
\expandafter\def\csname cnt@rev.desc.d3.sspro.p2.C.center-d1.reproduces-frozen\endcsname{1}
\expandafter\def\csname pct@rev.desc.d3.sspro.p2.C.center-d1.reproduces-frozen\endcsname{100.0}
\expandafter\def\csname ppt@rev.desc.d3.sspro.p2.C.center-d1.reproduces-frozen\endcsname{$+$100.0}
\expandafter\def\csname val@rev.desc.d3.sspro.p2.C.center-d1.resid-agree-stable\endcsname{0.024}
\expandafter\def\csname cnt@rev.desc.d3.sspro.p2.C.center-d1.resid-agree-stable\endcsname{0}
\expandafter\def\csname pct@rev.desc.d3.sspro.p2.C.center-d1.resid-agree-stable\endcsname{2.4}
\expandafter\def\csname ppt@rev.desc.d3.sspro.p2.C.center-d1.resid-agree-stable\endcsname{$+$2.4}
\expandafter\def\csname val@rev.desc.d3.sspro.p2.C.center-d1.resid-disagree\endcsname{0.000}
\expandafter\def\csname cnt@rev.desc.d3.sspro.p2.C.center-d1.resid-disagree\endcsname{0}
\expandafter\def\csname pct@rev.desc.d3.sspro.p2.C.center-d1.resid-disagree\endcsname{0.0}
\expandafter\def\csname ppt@rev.desc.d3.sspro.p2.C.center-d1.resid-disagree\endcsname{0.0}
\expandafter\def\csname val@rev.desc.d3.sspro.p2.C.center-d1.share-agree-stable\endcsname{0.875}
\expandafter\def\csname cnt@rev.desc.d3.sspro.p2.C.center-d1.share-agree-stable\endcsname{1}
\expandafter\def\csname pct@rev.desc.d3.sspro.p2.C.center-d1.share-agree-stable\endcsname{87.5}
\expandafter\def\csname ppt@rev.desc.d3.sspro.p2.C.center-d1.share-agree-stable\endcsname{$+$87.5}
\expandafter\def\csname val@rev.desc.d3.sspro.p2.C.context-d1.N-agree-stable\endcsname{40}
\expandafter\def\csname cnt@rev.desc.d3.sspro.p2.C.context-d1.N-agree-stable\endcsname{40}
\expandafter\def\csname pct@rev.desc.d3.sspro.p2.C.context-d1.N-agree-stable\endcsname{4000.0}
\expandafter\def\csname ppt@rev.desc.d3.sspro.p2.C.context-d1.N-agree-stable\endcsname{$+$4000.0}
\expandafter\def\csname val@rev.desc.d3.sspro.p2.C.context-d1.N-disagree\endcsname{7}
\expandafter\def\csname cnt@rev.desc.d3.sspro.p2.C.context-d1.N-disagree\endcsname{7}
\expandafter\def\csname pct@rev.desc.d3.sspro.p2.C.context-d1.N-disagree\endcsname{700.0}
\expandafter\def\csname ppt@rev.desc.d3.sspro.p2.C.context-d1.N-disagree\endcsname{$+$700.0}
\expandafter\def\csname val@rev.desc.d3.sspro.p2.C.context-d1.boot-resid.ci95.0\endcsname{0.000}
\expandafter\def\csname cnt@rev.desc.d3.sspro.p2.C.context-d1.boot-resid.ci95.0\endcsname{0}
\expandafter\def\csname pct@rev.desc.d3.sspro.p2.C.context-d1.boot-resid.ci95.0\endcsname{0.0}
\expandafter\def\csname ppt@rev.desc.d3.sspro.p2.C.context-d1.boot-resid.ci95.0\endcsname{0.0}
\expandafter\def\csname val@rev.desc.d3.sspro.p2.C.context-d1.boot-resid.ci95.1\endcsname{0.000}
\expandafter\def\csname cnt@rev.desc.d3.sspro.p2.C.context-d1.boot-resid.ci95.1\endcsname{0}
\expandafter\def\csname pct@rev.desc.d3.sspro.p2.C.context-d1.boot-resid.ci95.1\endcsname{0.0}
\expandafter\def\csname ppt@rev.desc.d3.sspro.p2.C.context-d1.boot-resid.ci95.1\endcsname{0.0}
\expandafter\def\csname val@rev.desc.d3.sspro.p2.C.context-d1.boot-resid.n-apps\endcsname{6}
\expandafter\def\csname cnt@rev.desc.d3.sspro.p2.C.context-d1.boot-resid.n-apps\endcsname{6}
\expandafter\def\csname pct@rev.desc.d3.sspro.p2.C.context-d1.boot-resid.n-apps\endcsname{600.0}
\expandafter\def\csname ppt@rev.desc.d3.sspro.p2.C.context-d1.boot-resid.n-apps\endcsname{$+$600.0}
\expandafter\def\csname val@rev.desc.d3.sspro.p2.C.context-d1.boot-resid.n-boot-valid\endcsname{4000}
\expandafter\def\csname cnt@rev.desc.d3.sspro.p2.C.context-d1.boot-resid.n-boot-valid\endcsname{4{,}000}
\expandafter\def\csname pct@rev.desc.d3.sspro.p2.C.context-d1.boot-resid.n-boot-valid\endcsname{400000.0}
\expandafter\def\csname ppt@rev.desc.d3.sspro.p2.C.context-d1.boot-resid.n-boot-valid\endcsname{$+$400000.0}
\expandafter\def\csname val@rev.desc.d3.sspro.p2.C.context-d1.boot-resid.value\endcsname{0.000}
\expandafter\def\csname cnt@rev.desc.d3.sspro.p2.C.context-d1.boot-resid.value\endcsname{0}
\expandafter\def\csname pct@rev.desc.d3.sspro.p2.C.context-d1.boot-resid.value\endcsname{0.0}
\expandafter\def\csname ppt@rev.desc.d3.sspro.p2.C.context-d1.boot-resid.value\endcsname{0.0}
\expandafter\def\csname val@rev.desc.d3.sspro.p2.C.context-d1.boot-share.ci95.0\endcsname{0.583}
\expandafter\def\csname cnt@rev.desc.d3.sspro.p2.C.context-d1.boot-share.ci95.0\endcsname{1}
\expandafter\def\csname pct@rev.desc.d3.sspro.p2.C.context-d1.boot-share.ci95.0\endcsname{58.3}
\expandafter\def\csname ppt@rev.desc.d3.sspro.p2.C.context-d1.boot-share.ci95.0\endcsname{$+$58.3}
\expandafter\def\csname val@rev.desc.d3.sspro.p2.C.context-d1.boot-share.ci95.1\endcsname{0.979}
\expandafter\def\csname cnt@rev.desc.d3.sspro.p2.C.context-d1.boot-share.ci95.1\endcsname{1}
\expandafter\def\csname pct@rev.desc.d3.sspro.p2.C.context-d1.boot-share.ci95.1\endcsname{97.9}
\expandafter\def\csname ppt@rev.desc.d3.sspro.p2.C.context-d1.boot-share.ci95.1\endcsname{$+$97.9}
\expandafter\def\csname val@rev.desc.d3.sspro.p2.C.context-d1.boot-share.n-apps\endcsname{6}
\expandafter\def\csname cnt@rev.desc.d3.sspro.p2.C.context-d1.boot-share.n-apps\endcsname{6}
\expandafter\def\csname pct@rev.desc.d3.sspro.p2.C.context-d1.boot-share.n-apps\endcsname{600.0}
\expandafter\def\csname ppt@rev.desc.d3.sspro.p2.C.context-d1.boot-share.n-apps\endcsname{$+$600.0}
\expandafter\def\csname val@rev.desc.d3.sspro.p2.C.context-d1.boot-share.n-boot-valid\endcsname{4000}
\expandafter\def\csname cnt@rev.desc.d3.sspro.p2.C.context-d1.boot-share.n-boot-valid\endcsname{4{,}000}
\expandafter\def\csname pct@rev.desc.d3.sspro.p2.C.context-d1.boot-share.n-boot-valid\endcsname{400000.0}
\expandafter\def\csname ppt@rev.desc.d3.sspro.p2.C.context-d1.boot-share.n-boot-valid\endcsname{$+$400000.0}
\expandafter\def\csname val@rev.desc.d3.sspro.p2.C.context-d1.boot-share.value\endcsname{0.833}
\expandafter\def\csname cnt@rev.desc.d3.sspro.p2.C.context-d1.boot-share.value\endcsname{1}
\expandafter\def\csname pct@rev.desc.d3.sspro.p2.C.context-d1.boot-share.value\endcsname{83.3}
\expandafter\def\csname ppt@rev.desc.d3.sspro.p2.C.context-d1.boot-share.value\endcsname{$+$83.3}
\expandafter\def\csname val@rev.desc.d3.sspro.p2.C.context-d1.n\endcsname{48}
\expandafter\def\csname cnt@rev.desc.d3.sspro.p2.C.context-d1.n\endcsname{48}
\expandafter\def\csname pct@rev.desc.d3.sspro.p2.C.context-d1.n\endcsname{4800.0}
\expandafter\def\csname ppt@rev.desc.d3.sspro.p2.C.context-d1.n\endcsname{$+$4800.0}
\expandafter\def\csname val@rev.desc.d3.sspro.p2.C.context-d1.reproduces-frozen\endcsname{1}
\expandafter\def\csname cnt@rev.desc.d3.sspro.p2.C.context-d1.reproduces-frozen\endcsname{1}
\expandafter\def\csname pct@rev.desc.d3.sspro.p2.C.context-d1.reproduces-frozen\endcsname{100.0}
\expandafter\def\csname ppt@rev.desc.d3.sspro.p2.C.context-d1.reproduces-frozen\endcsname{$+$100.0}
\expandafter\def\csname val@rev.desc.d3.sspro.p2.C.context-d1.resid-agree-stable\endcsname{0.000}
\expandafter\def\csname cnt@rev.desc.d3.sspro.p2.C.context-d1.resid-agree-stable\endcsname{0}
\expandafter\def\csname pct@rev.desc.d3.sspro.p2.C.context-d1.resid-agree-stable\endcsname{0.0}
\expandafter\def\csname ppt@rev.desc.d3.sspro.p2.C.context-d1.resid-agree-stable\endcsname{0.0}
\expandafter\def\csname val@rev.desc.d3.sspro.p2.C.context-d1.resid-disagree\endcsname{0.000}
\expandafter\def\csname cnt@rev.desc.d3.sspro.p2.C.context-d1.resid-disagree\endcsname{0}
\expandafter\def\csname pct@rev.desc.d3.sspro.p2.C.context-d1.resid-disagree\endcsname{0.0}
\expandafter\def\csname ppt@rev.desc.d3.sspro.p2.C.context-d1.resid-disagree\endcsname{0.0}
\expandafter\def\csname val@rev.desc.d3.sspro.p2.C.context-d1.share-agree-stable\endcsname{0.833}
\expandafter\def\csname cnt@rev.desc.d3.sspro.p2.C.context-d1.share-agree-stable\endcsname{1}
\expandafter\def\csname pct@rev.desc.d3.sspro.p2.C.context-d1.share-agree-stable\endcsname{83.3}
\expandafter\def\csname ppt@rev.desc.d3.sspro.p2.C.context-d1.share-agree-stable\endcsname{$+$83.3}
\expandafter\def\csname val@rev.desc.d3.sspro.p2.C.dual-d2.N-agree-stable\endcsname{39}
\expandafter\def\csname cnt@rev.desc.d3.sspro.p2.C.dual-d2.N-agree-stable\endcsname{39}
\expandafter\def\csname pct@rev.desc.d3.sspro.p2.C.dual-d2.N-agree-stable\endcsname{3900.0}
\expandafter\def\csname ppt@rev.desc.d3.sspro.p2.C.dual-d2.N-agree-stable\endcsname{$+$3900.0}
\expandafter\def\csname val@rev.desc.d3.sspro.p2.C.dual-d2.N-disagree\endcsname{9}
\expandafter\def\csname cnt@rev.desc.d3.sspro.p2.C.dual-d2.N-disagree\endcsname{9}
\expandafter\def\csname pct@rev.desc.d3.sspro.p2.C.dual-d2.N-disagree\endcsname{900.0}
\expandafter\def\csname ppt@rev.desc.d3.sspro.p2.C.dual-d2.N-disagree\endcsname{$+$900.0}
\expandafter\def\csname val@rev.desc.d3.sspro.p2.C.dual-d2.boot-resid.ci95.0\endcsname{-0.179}
\expandafter\def\csname cnt@rev.desc.d3.sspro.p2.C.dual-d2.boot-resid.ci95.0\endcsname{0}
\expandafter\def\csname pct@rev.desc.d3.sspro.p2.C.dual-d2.boot-resid.ci95.0\endcsname{$-$17.9}
\expandafter\def\csname ppt@rev.desc.d3.sspro.p2.C.dual-d2.boot-resid.ci95.0\endcsname{$-$17.9}
\expandafter\def\csname val@rev.desc.d3.sspro.p2.C.dual-d2.boot-resid.ci95.1\endcsname{0.048}
\expandafter\def\csname cnt@rev.desc.d3.sspro.p2.C.dual-d2.boot-resid.ci95.1\endcsname{0}
\expandafter\def\csname pct@rev.desc.d3.sspro.p2.C.dual-d2.boot-resid.ci95.1\endcsname{4.8}
\expandafter\def\csname ppt@rev.desc.d3.sspro.p2.C.dual-d2.boot-resid.ci95.1\endcsname{$+$4.8}
\expandafter\def\csname val@rev.desc.d3.sspro.p2.C.dual-d2.boot-resid.n-apps\endcsname{6}
\expandafter\def\csname cnt@rev.desc.d3.sspro.p2.C.dual-d2.boot-resid.n-apps\endcsname{6}
\expandafter\def\csname pct@rev.desc.d3.sspro.p2.C.dual-d2.boot-resid.n-apps\endcsname{600.0}
\expandafter\def\csname ppt@rev.desc.d3.sspro.p2.C.dual-d2.boot-resid.n-apps\endcsname{$+$600.0}
\expandafter\def\csname val@rev.desc.d3.sspro.p2.C.dual-d2.boot-resid.n-boot-valid\endcsname{4000}
\expandafter\def\csname cnt@rev.desc.d3.sspro.p2.C.dual-d2.boot-resid.n-boot-valid\endcsname{4{,}000}
\expandafter\def\csname pct@rev.desc.d3.sspro.p2.C.dual-d2.boot-resid.n-boot-valid\endcsname{400000.0}
\expandafter\def\csname ppt@rev.desc.d3.sspro.p2.C.dual-d2.boot-resid.n-boot-valid\endcsname{$+$400000.0}
\expandafter\def\csname val@rev.desc.d3.sspro.p2.C.dual-d2.boot-resid.value\endcsname{-0.051}
\expandafter\def\csname cnt@rev.desc.d3.sspro.p2.C.dual-d2.boot-resid.value\endcsname{0}
\expandafter\def\csname pct@rev.desc.d3.sspro.p2.C.dual-d2.boot-resid.value\endcsname{$-$5.1}
\expandafter\def\csname ppt@rev.desc.d3.sspro.p2.C.dual-d2.boot-resid.value\endcsname{$-$5.1}
\expandafter\def\csname val@rev.desc.d3.sspro.p2.C.dual-d2.boot-share.ci95.0\endcsname{0.562}
\expandafter\def\csname cnt@rev.desc.d3.sspro.p2.C.dual-d2.boot-share.ci95.0\endcsname{1}
\expandafter\def\csname pct@rev.desc.d3.sspro.p2.C.dual-d2.boot-share.ci95.0\endcsname{56.2}
\expandafter\def\csname ppt@rev.desc.d3.sspro.p2.C.dual-d2.boot-share.ci95.0\endcsname{$+$56.2}
\expandafter\def\csname val@rev.desc.d3.sspro.p2.C.dual-d2.boot-share.ci95.1\endcsname{0.958}
\expandafter\def\csname cnt@rev.desc.d3.sspro.p2.C.dual-d2.boot-share.ci95.1\endcsname{1}
\expandafter\def\csname pct@rev.desc.d3.sspro.p2.C.dual-d2.boot-share.ci95.1\endcsname{95.8}
\expandafter\def\csname ppt@rev.desc.d3.sspro.p2.C.dual-d2.boot-share.ci95.1\endcsname{$+$95.8}
\expandafter\def\csname val@rev.desc.d3.sspro.p2.C.dual-d2.boot-share.n-apps\endcsname{6}
\expandafter\def\csname cnt@rev.desc.d3.sspro.p2.C.dual-d2.boot-share.n-apps\endcsname{6}
\expandafter\def\csname pct@rev.desc.d3.sspro.p2.C.dual-d2.boot-share.n-apps\endcsname{600.0}
\expandafter\def\csname ppt@rev.desc.d3.sspro.p2.C.dual-d2.boot-share.n-apps\endcsname{$+$600.0}
\expandafter\def\csname val@rev.desc.d3.sspro.p2.C.dual-d2.boot-share.n-boot-valid\endcsname{4000}
\expandafter\def\csname cnt@rev.desc.d3.sspro.p2.C.dual-d2.boot-share.n-boot-valid\endcsname{4{,}000}
\expandafter\def\csname pct@rev.desc.d3.sspro.p2.C.dual-d2.boot-share.n-boot-valid\endcsname{400000.0}
\expandafter\def\csname ppt@rev.desc.d3.sspro.p2.C.dual-d2.boot-share.n-boot-valid\endcsname{$+$400000.0}
\expandafter\def\csname val@rev.desc.d3.sspro.p2.C.dual-d2.boot-share.value\endcsname{0.812}
\expandafter\def\csname cnt@rev.desc.d3.sspro.p2.C.dual-d2.boot-share.value\endcsname{1}
\expandafter\def\csname pct@rev.desc.d3.sspro.p2.C.dual-d2.boot-share.value\endcsname{81.2}
\expandafter\def\csname ppt@rev.desc.d3.sspro.p2.C.dual-d2.boot-share.value\endcsname{$+$81.2}
\expandafter\def\csname val@rev.desc.d3.sspro.p2.C.dual-d2.n\endcsname{48}
\expandafter\def\csname cnt@rev.desc.d3.sspro.p2.C.dual-d2.n\endcsname{48}
\expandafter\def\csname pct@rev.desc.d3.sspro.p2.C.dual-d2.n\endcsname{4800.0}
\expandafter\def\csname ppt@rev.desc.d3.sspro.p2.C.dual-d2.n\endcsname{$+$4800.0}
\expandafter\def\csname val@rev.desc.d3.sspro.p2.C.dual-d2.reproduces-frozen\endcsname{1}
\expandafter\def\csname cnt@rev.desc.d3.sspro.p2.C.dual-d2.reproduces-frozen\endcsname{1}
\expandafter\def\csname pct@rev.desc.d3.sspro.p2.C.dual-d2.reproduces-frozen\endcsname{100.0}
\expandafter\def\csname ppt@rev.desc.d3.sspro.p2.C.dual-d2.reproduces-frozen\endcsname{$+$100.0}
\expandafter\def\csname val@rev.desc.d3.sspro.p2.C.dual-d2.resid-agree-stable\endcsname{-0.051}
\expandafter\def\csname cnt@rev.desc.d3.sspro.p2.C.dual-d2.resid-agree-stable\endcsname{0}
\expandafter\def\csname pct@rev.desc.d3.sspro.p2.C.dual-d2.resid-agree-stable\endcsname{$-$5.1}
\expandafter\def\csname ppt@rev.desc.d3.sspro.p2.C.dual-d2.resid-agree-stable\endcsname{$-$5.1}
\expandafter\def\csname val@rev.desc.d3.sspro.p2.C.dual-d2.resid-disagree\endcsname{0.333}
\expandafter\def\csname cnt@rev.desc.d3.sspro.p2.C.dual-d2.resid-disagree\endcsname{0}
\expandafter\def\csname pct@rev.desc.d3.sspro.p2.C.dual-d2.resid-disagree\endcsname{33.3}
\expandafter\def\csname ppt@rev.desc.d3.sspro.p2.C.dual-d2.resid-disagree\endcsname{$+$33.3}
\expandafter\def\csname val@rev.desc.d3.sspro.p2.C.dual-d2.share-agree-stable\endcsname{0.812}
\expandafter\def\csname cnt@rev.desc.d3.sspro.p2.C.dual-d2.share-agree-stable\endcsname{1}
\expandafter\def\csname pct@rev.desc.d3.sspro.p2.C.dual-d2.share-agree-stable\endcsname{81.2}
\expandafter\def\csname ppt@rev.desc.d3.sspro.p2.C.dual-d2.share-agree-stable\endcsname{$+$81.2}
\expandafter\def\csname val@rev.desc.d3.sspro.p2.C.grid2-d3.N-agree-stable\endcsname{39}
\expandafter\def\csname cnt@rev.desc.d3.sspro.p2.C.grid2-d3.N-agree-stable\endcsname{39}
\expandafter\def\csname pct@rev.desc.d3.sspro.p2.C.grid2-d3.N-agree-stable\endcsname{3900.0}
\expandafter\def\csname ppt@rev.desc.d3.sspro.p2.C.grid2-d3.N-agree-stable\endcsname{$+$3900.0}
\expandafter\def\csname val@rev.desc.d3.sspro.p2.C.grid2-d3.N-disagree\endcsname{6}
\expandafter\def\csname cnt@rev.desc.d3.sspro.p2.C.grid2-d3.N-disagree\endcsname{6}
\expandafter\def\csname pct@rev.desc.d3.sspro.p2.C.grid2-d3.N-disagree\endcsname{600.0}
\expandafter\def\csname ppt@rev.desc.d3.sspro.p2.C.grid2-d3.N-disagree\endcsname{$+$600.0}
\expandafter\def\csname val@rev.desc.d3.sspro.p2.C.grid2-d3.boot-resid.ci95.0\endcsname{-0.073}
\expandafter\def\csname cnt@rev.desc.d3.sspro.p2.C.grid2-d3.boot-resid.ci95.0\endcsname{0}
\expandafter\def\csname pct@rev.desc.d3.sspro.p2.C.grid2-d3.boot-resid.ci95.0\endcsname{$-$7.3}
\expandafter\def\csname ppt@rev.desc.d3.sspro.p2.C.grid2-d3.boot-resid.ci95.0\endcsname{$-$7.3}
\expandafter\def\csname val@rev.desc.d3.sspro.p2.C.grid2-d3.boot-resid.ci95.1\endcsname{0.000}
\expandafter\def\csname cnt@rev.desc.d3.sspro.p2.C.grid2-d3.boot-resid.ci95.1\endcsname{0}
\expandafter\def\csname pct@rev.desc.d3.sspro.p2.C.grid2-d3.boot-resid.ci95.1\endcsname{0.0}
\expandafter\def\csname ppt@rev.desc.d3.sspro.p2.C.grid2-d3.boot-resid.ci95.1\endcsname{0.0}
\expandafter\def\csname val@rev.desc.d3.sspro.p2.C.grid2-d3.boot-resid.n-apps\endcsname{6}
\expandafter\def\csname cnt@rev.desc.d3.sspro.p2.C.grid2-d3.boot-resid.n-apps\endcsname{6}
\expandafter\def\csname pct@rev.desc.d3.sspro.p2.C.grid2-d3.boot-resid.n-apps\endcsname{600.0}
\expandafter\def\csname ppt@rev.desc.d3.sspro.p2.C.grid2-d3.boot-resid.n-apps\endcsname{$+$600.0}
\expandafter\def\csname val@rev.desc.d3.sspro.p2.C.grid2-d3.boot-resid.n-boot-valid\endcsname{4000}
\expandafter\def\csname cnt@rev.desc.d3.sspro.p2.C.grid2-d3.boot-resid.n-boot-valid\endcsname{4{,}000}
\expandafter\def\csname pct@rev.desc.d3.sspro.p2.C.grid2-d3.boot-resid.n-boot-valid\endcsname{400000.0}
\expandafter\def\csname ppt@rev.desc.d3.sspro.p2.C.grid2-d3.boot-resid.n-boot-valid\endcsname{$+$400000.0}
\expandafter\def\csname val@rev.desc.d3.sspro.p2.C.grid2-d3.boot-resid.value\endcsname{-0.026}
\expandafter\def\csname cnt@rev.desc.d3.sspro.p2.C.grid2-d3.boot-resid.value\endcsname{0}
\expandafter\def\csname pct@rev.desc.d3.sspro.p2.C.grid2-d3.boot-resid.value\endcsname{$-$2.6}
\expandafter\def\csname ppt@rev.desc.d3.sspro.p2.C.grid2-d3.boot-resid.value\endcsname{$-$2.6}
\expandafter\def\csname val@rev.desc.d3.sspro.p2.C.grid2-d3.boot-share.ci95.0\endcsname{0.583}
\expandafter\def\csname cnt@rev.desc.d3.sspro.p2.C.grid2-d3.boot-share.ci95.0\endcsname{1}
\expandafter\def\csname pct@rev.desc.d3.sspro.p2.C.grid2-d3.boot-share.ci95.0\endcsname{58.3}
\expandafter\def\csname ppt@rev.desc.d3.sspro.p2.C.grid2-d3.boot-share.ci95.0\endcsname{$+$58.3}
\expandafter\def\csname val@rev.desc.d3.sspro.p2.C.grid2-d3.boot-share.ci95.1\endcsname{0.958}
\expandafter\def\csname cnt@rev.desc.d3.sspro.p2.C.grid2-d3.boot-share.ci95.1\endcsname{1}
\expandafter\def\csname pct@rev.desc.d3.sspro.p2.C.grid2-d3.boot-share.ci95.1\endcsname{95.8}
\expandafter\def\csname ppt@rev.desc.d3.sspro.p2.C.grid2-d3.boot-share.ci95.1\endcsname{$+$95.8}
\expandafter\def\csname val@rev.desc.d3.sspro.p2.C.grid2-d3.boot-share.n-apps\endcsname{6}
\expandafter\def\csname cnt@rev.desc.d3.sspro.p2.C.grid2-d3.boot-share.n-apps\endcsname{6}
\expandafter\def\csname pct@rev.desc.d3.sspro.p2.C.grid2-d3.boot-share.n-apps\endcsname{600.0}
\expandafter\def\csname ppt@rev.desc.d3.sspro.p2.C.grid2-d3.boot-share.n-apps\endcsname{$+$600.0}
\expandafter\def\csname val@rev.desc.d3.sspro.p2.C.grid2-d3.boot-share.n-boot-valid\endcsname{4000}
\expandafter\def\csname cnt@rev.desc.d3.sspro.p2.C.grid2-d3.boot-share.n-boot-valid\endcsname{4{,}000}
\expandafter\def\csname pct@rev.desc.d3.sspro.p2.C.grid2-d3.boot-share.n-boot-valid\endcsname{400000.0}
\expandafter\def\csname ppt@rev.desc.d3.sspro.p2.C.grid2-d3.boot-share.n-boot-valid\endcsname{$+$400000.0}
\expandafter\def\csname val@rev.desc.d3.sspro.p2.C.grid2-d3.boot-share.value\endcsname{0.812}
\expandafter\def\csname cnt@rev.desc.d3.sspro.p2.C.grid2-d3.boot-share.value\endcsname{1}
\expandafter\def\csname pct@rev.desc.d3.sspro.p2.C.grid2-d3.boot-share.value\endcsname{81.2}
\expandafter\def\csname ppt@rev.desc.d3.sspro.p2.C.grid2-d3.boot-share.value\endcsname{$+$81.2}
\expandafter\def\csname val@rev.desc.d3.sspro.p2.C.grid2-d3.n\endcsname{48}
\expandafter\def\csname cnt@rev.desc.d3.sspro.p2.C.grid2-d3.n\endcsname{48}
\expandafter\def\csname pct@rev.desc.d3.sspro.p2.C.grid2-d3.n\endcsname{4800.0}
\expandafter\def\csname ppt@rev.desc.d3.sspro.p2.C.grid2-d3.n\endcsname{$+$4800.0}
\expandafter\def\csname val@rev.desc.d3.sspro.p2.C.grid2-d3.reproduces-frozen\endcsname{1}
\expandafter\def\csname cnt@rev.desc.d3.sspro.p2.C.grid2-d3.reproduces-frozen\endcsname{1}
\expandafter\def\csname pct@rev.desc.d3.sspro.p2.C.grid2-d3.reproduces-frozen\endcsname{100.0}
\expandafter\def\csname ppt@rev.desc.d3.sspro.p2.C.grid2-d3.reproduces-frozen\endcsname{$+$100.0}
\expandafter\def\csname val@rev.desc.d3.sspro.p2.C.grid2-d3.resid-agree-stable\endcsname{-0.026}
\expandafter\def\csname cnt@rev.desc.d3.sspro.p2.C.grid2-d3.resid-agree-stable\endcsname{0}
\expandafter\def\csname pct@rev.desc.d3.sspro.p2.C.grid2-d3.resid-agree-stable\endcsname{$-$2.6}
\expandafter\def\csname ppt@rev.desc.d3.sspro.p2.C.grid2-d3.resid-agree-stable\endcsname{$-$2.6}
\expandafter\def\csname val@rev.desc.d3.sspro.p2.C.grid2-d3.resid-disagree\endcsname{0.000}
\expandafter\def\csname cnt@rev.desc.d3.sspro.p2.C.grid2-d3.resid-disagree\endcsname{0}
\expandafter\def\csname pct@rev.desc.d3.sspro.p2.C.grid2-d3.resid-disagree\endcsname{0.0}
\expandafter\def\csname ppt@rev.desc.d3.sspro.p2.C.grid2-d3.resid-disagree\endcsname{0.0}
\expandafter\def\csname val@rev.desc.d3.sspro.p2.C.grid2-d3.share-agree-stable\endcsname{0.812}
\expandafter\def\csname cnt@rev.desc.d3.sspro.p2.C.grid2-d3.share-agree-stable\endcsname{1}
\expandafter\def\csname pct@rev.desc.d3.sspro.p2.C.grid2-d3.share-agree-stable\endcsname{81.2}
\expandafter\def\csname ppt@rev.desc.d3.sspro.p2.C.grid2-d3.share-agree-stable\endcsname{$+$81.2}
\expandafter\def\csname val@rev.desc.d3.sspro.p2.C.zoomclick-d3.N-agree-stable\endcsname{41}
\expandafter\def\csname cnt@rev.desc.d3.sspro.p2.C.zoomclick-d3.N-agree-stable\endcsname{41}
\expandafter\def\csname pct@rev.desc.d3.sspro.p2.C.zoomclick-d3.N-agree-stable\endcsname{4100.0}
\expandafter\def\csname ppt@rev.desc.d3.sspro.p2.C.zoomclick-d3.N-agree-stable\endcsname{$+$4100.0}
\expandafter\def\csname val@rev.desc.d3.sspro.p2.C.zoomclick-d3.N-disagree\endcsname{6}
\expandafter\def\csname cnt@rev.desc.d3.sspro.p2.C.zoomclick-d3.N-disagree\endcsname{6}
\expandafter\def\csname pct@rev.desc.d3.sspro.p2.C.zoomclick-d3.N-disagree\endcsname{600.0}
\expandafter\def\csname ppt@rev.desc.d3.sspro.p2.C.zoomclick-d3.N-disagree\endcsname{$+$600.0}
\expandafter\def\csname val@rev.desc.d3.sspro.p2.C.zoomclick-d3.boot-resid.ci95.0\endcsname{0.000}
\expandafter\def\csname cnt@rev.desc.d3.sspro.p2.C.zoomclick-d3.boot-resid.ci95.0\endcsname{0}
\expandafter\def\csname pct@rev.desc.d3.sspro.p2.C.zoomclick-d3.boot-resid.ci95.0\endcsname{0.0}
\expandafter\def\csname ppt@rev.desc.d3.sspro.p2.C.zoomclick-d3.boot-resid.ci95.0\endcsname{0.0}
\expandafter\def\csname val@rev.desc.d3.sspro.p2.C.zoomclick-d3.boot-resid.ci95.1\endcsname{0.000}
\expandafter\def\csname cnt@rev.desc.d3.sspro.p2.C.zoomclick-d3.boot-resid.ci95.1\endcsname{0}
\expandafter\def\csname pct@rev.desc.d3.sspro.p2.C.zoomclick-d3.boot-resid.ci95.1\endcsname{0.0}
\expandafter\def\csname ppt@rev.desc.d3.sspro.p2.C.zoomclick-d3.boot-resid.ci95.1\endcsname{0.0}
\expandafter\def\csname val@rev.desc.d3.sspro.p2.C.zoomclick-d3.boot-resid.n-apps\endcsname{6}
\expandafter\def\csname cnt@rev.desc.d3.sspro.p2.C.zoomclick-d3.boot-resid.n-apps\endcsname{6}
\expandafter\def\csname pct@rev.desc.d3.sspro.p2.C.zoomclick-d3.boot-resid.n-apps\endcsname{600.0}
\expandafter\def\csname ppt@rev.desc.d3.sspro.p2.C.zoomclick-d3.boot-resid.n-apps\endcsname{$+$600.0}
\expandafter\def\csname val@rev.desc.d3.sspro.p2.C.zoomclick-d3.boot-resid.n-boot-valid\endcsname{4000}
\expandafter\def\csname cnt@rev.desc.d3.sspro.p2.C.zoomclick-d3.boot-resid.n-boot-valid\endcsname{4{,}000}
\expandafter\def\csname pct@rev.desc.d3.sspro.p2.C.zoomclick-d3.boot-resid.n-boot-valid\endcsname{400000.0}
\expandafter\def\csname ppt@rev.desc.d3.sspro.p2.C.zoomclick-d3.boot-resid.n-boot-valid\endcsname{$+$400000.0}
\expandafter\def\csname val@rev.desc.d3.sspro.p2.C.zoomclick-d3.boot-resid.value\endcsname{0.000}
\expandafter\def\csname cnt@rev.desc.d3.sspro.p2.C.zoomclick-d3.boot-resid.value\endcsname{0}
\expandafter\def\csname pct@rev.desc.d3.sspro.p2.C.zoomclick-d3.boot-resid.value\endcsname{0.0}
\expandafter\def\csname ppt@rev.desc.d3.sspro.p2.C.zoomclick-d3.boot-resid.value\endcsname{0.0}
\expandafter\def\csname val@rev.desc.d3.sspro.p2.C.zoomclick-d3.boot-share.ci95.0\endcsname{0.604}
\expandafter\def\csname cnt@rev.desc.d3.sspro.p2.C.zoomclick-d3.boot-share.ci95.0\endcsname{1}
\expandafter\def\csname pct@rev.desc.d3.sspro.p2.C.zoomclick-d3.boot-share.ci95.0\endcsname{60.4}
\expandafter\def\csname ppt@rev.desc.d3.sspro.p2.C.zoomclick-d3.boot-share.ci95.0\endcsname{$+$60.4}
\expandafter\def\csname val@rev.desc.d3.sspro.p2.C.zoomclick-d3.boot-share.ci95.1\endcsname{1.000}
\expandafter\def\csname cnt@rev.desc.d3.sspro.p2.C.zoomclick-d3.boot-share.ci95.1\endcsname{1}
\expandafter\def\csname pct@rev.desc.d3.sspro.p2.C.zoomclick-d3.boot-share.ci95.1\endcsname{100.0}
\expandafter\def\csname ppt@rev.desc.d3.sspro.p2.C.zoomclick-d3.boot-share.ci95.1\endcsname{$+$100.0}
\expandafter\def\csname val@rev.desc.d3.sspro.p2.C.zoomclick-d3.boot-share.n-apps\endcsname{6}
\expandafter\def\csname cnt@rev.desc.d3.sspro.p2.C.zoomclick-d3.boot-share.n-apps\endcsname{6}
\expandafter\def\csname pct@rev.desc.d3.sspro.p2.C.zoomclick-d3.boot-share.n-apps\endcsname{600.0}
\expandafter\def\csname ppt@rev.desc.d3.sspro.p2.C.zoomclick-d3.boot-share.n-apps\endcsname{$+$600.0}
\expandafter\def\csname val@rev.desc.d3.sspro.p2.C.zoomclick-d3.boot-share.n-boot-valid\endcsname{4000}
\expandafter\def\csname cnt@rev.desc.d3.sspro.p2.C.zoomclick-d3.boot-share.n-boot-valid\endcsname{4{,}000}
\expandafter\def\csname pct@rev.desc.d3.sspro.p2.C.zoomclick-d3.boot-share.n-boot-valid\endcsname{400000.0}
\expandafter\def\csname ppt@rev.desc.d3.sspro.p2.C.zoomclick-d3.boot-share.n-boot-valid\endcsname{$+$400000.0}
\expandafter\def\csname val@rev.desc.d3.sspro.p2.C.zoomclick-d3.boot-share.value\endcsname{0.854}
\expandafter\def\csname cnt@rev.desc.d3.sspro.p2.C.zoomclick-d3.boot-share.value\endcsname{1}
\expandafter\def\csname pct@rev.desc.d3.sspro.p2.C.zoomclick-d3.boot-share.value\endcsname{85.4}
\expandafter\def\csname ppt@rev.desc.d3.sspro.p2.C.zoomclick-d3.boot-share.value\endcsname{$+$85.4}
\expandafter\def\csname val@rev.desc.d3.sspro.p2.C.zoomclick-d3.n\endcsname{48}
\expandafter\def\csname cnt@rev.desc.d3.sspro.p2.C.zoomclick-d3.n\endcsname{48}
\expandafter\def\csname pct@rev.desc.d3.sspro.p2.C.zoomclick-d3.n\endcsname{4800.0}
\expandafter\def\csname ppt@rev.desc.d3.sspro.p2.C.zoomclick-d3.n\endcsname{$+$4800.0}
\expandafter\def\csname val@rev.desc.d3.sspro.p2.C.zoomclick-d3.reproduces-frozen\endcsname{1}
\expandafter\def\csname cnt@rev.desc.d3.sspro.p2.C.zoomclick-d3.reproduces-frozen\endcsname{1}
\expandafter\def\csname pct@rev.desc.d3.sspro.p2.C.zoomclick-d3.reproduces-frozen\endcsname{100.0}
\expandafter\def\csname ppt@rev.desc.d3.sspro.p2.C.zoomclick-d3.reproduces-frozen\endcsname{$+$100.0}
\expandafter\def\csname val@rev.desc.d3.sspro.p2.C.zoomclick-d3.resid-agree-stable\endcsname{0.000}
\expandafter\def\csname cnt@rev.desc.d3.sspro.p2.C.zoomclick-d3.resid-agree-stable\endcsname{0}
\expandafter\def\csname pct@rev.desc.d3.sspro.p2.C.zoomclick-d3.resid-agree-stable\endcsname{0.0}
\expandafter\def\csname ppt@rev.desc.d3.sspro.p2.C.zoomclick-d3.resid-agree-stable\endcsname{0.0}
\expandafter\def\csname val@rev.desc.d3.sspro.p2.C.zoomclick-d3.resid-disagree\endcsname{0.333}
\expandafter\def\csname cnt@rev.desc.d3.sspro.p2.C.zoomclick-d3.resid-disagree\endcsname{0}
\expandafter\def\csname pct@rev.desc.d3.sspro.p2.C.zoomclick-d3.resid-disagree\endcsname{33.3}
\expandafter\def\csname ppt@rev.desc.d3.sspro.p2.C.zoomclick-d3.resid-disagree\endcsname{$+$33.3}
\expandafter\def\csname val@rev.desc.d3.sspro.p2.C.zoomclick-d3.share-agree-stable\endcsname{0.854}
\expandafter\def\csname cnt@rev.desc.d3.sspro.p2.C.zoomclick-d3.share-agree-stable\endcsname{1}
\expandafter\def\csname pct@rev.desc.d3.sspro.p2.C.zoomclick-d3.share-agree-stable\endcsname{85.4}
\expandafter\def\csname ppt@rev.desc.d3.sspro.p2.C.zoomclick-d3.share-agree-stable\endcsname{$+$85.4}
\expandafter\def\csname val@rev.desc.d3.sspro.p2.T.catalog-pooled.N-agree-stable\endcsname{334}
\expandafter\def\csname cnt@rev.desc.d3.sspro.p2.T.catalog-pooled.N-agree-stable\endcsname{334}
\expandafter\def\csname pct@rev.desc.d3.sspro.p2.T.catalog-pooled.N-agree-stable\endcsname{33400.0}
\expandafter\def\csname ppt@rev.desc.d3.sspro.p2.T.catalog-pooled.N-agree-stable\endcsname{$+$33400.0}
\expandafter\def\csname val@rev.desc.d3.sspro.p2.T.catalog-pooled.N-disagree\endcsname{41}
\expandafter\def\csname cnt@rev.desc.d3.sspro.p2.T.catalog-pooled.N-disagree\endcsname{41}
\expandafter\def\csname pct@rev.desc.d3.sspro.p2.T.catalog-pooled.N-disagree\endcsname{4100.0}
\expandafter\def\csname ppt@rev.desc.d3.sspro.p2.T.catalog-pooled.N-disagree\endcsname{$+$4100.0}
\expandafter\def\csname val@rev.desc.d3.sspro.p2.T.catalog-pooled.boot-resid.ci95.0\endcsname{-0.012}
\expandafter\def\csname cnt@rev.desc.d3.sspro.p2.T.catalog-pooled.boot-resid.ci95.0\endcsname{0}
\expandafter\def\csname pct@rev.desc.d3.sspro.p2.T.catalog-pooled.boot-resid.ci95.0\endcsname{$-$1.2}
\expandafter\def\csname ppt@rev.desc.d3.sspro.p2.T.catalog-pooled.boot-resid.ci95.0\endcsname{$-$1.2}
\expandafter\def\csname val@rev.desc.d3.sspro.p2.T.catalog-pooled.boot-resid.ci95.1\endcsname{0.028}
\expandafter\def\csname cnt@rev.desc.d3.sspro.p2.T.catalog-pooled.boot-resid.ci95.1\endcsname{0}
\expandafter\def\csname pct@rev.desc.d3.sspro.p2.T.catalog-pooled.boot-resid.ci95.1\endcsname{2.8}
\expandafter\def\csname ppt@rev.desc.d3.sspro.p2.T.catalog-pooled.boot-resid.ci95.1\endcsname{$+$2.8}
\expandafter\def\csname val@rev.desc.d3.sspro.p2.T.catalog-pooled.boot-resid.n-apps\endcsname{12}
\expandafter\def\csname cnt@rev.desc.d3.sspro.p2.T.catalog-pooled.boot-resid.n-apps\endcsname{12}
\expandafter\def\csname pct@rev.desc.d3.sspro.p2.T.catalog-pooled.boot-resid.n-apps\endcsname{1200.0}
\expandafter\def\csname ppt@rev.desc.d3.sspro.p2.T.catalog-pooled.boot-resid.n-apps\endcsname{$+$1200.0}
\expandafter\def\csname val@rev.desc.d3.sspro.p2.T.catalog-pooled.boot-resid.n-boot-valid\endcsname{4000}
\expandafter\def\csname cnt@rev.desc.d3.sspro.p2.T.catalog-pooled.boot-resid.n-boot-valid\endcsname{4{,}000}
\expandafter\def\csname pct@rev.desc.d3.sspro.p2.T.catalog-pooled.boot-resid.n-boot-valid\endcsname{400000.0}
\expandafter\def\csname ppt@rev.desc.d3.sspro.p2.T.catalog-pooled.boot-resid.n-boot-valid\endcsname{$+$400000.0}
\expandafter\def\csname val@rev.desc.d3.sspro.p2.T.catalog-pooled.boot-resid.value\endcsname{0.009}
\expandafter\def\csname cnt@rev.desc.d3.sspro.p2.T.catalog-pooled.boot-resid.value\endcsname{0}
\expandafter\def\csname pct@rev.desc.d3.sspro.p2.T.catalog-pooled.boot-resid.value\endcsname{0.9}
\expandafter\def\csname ppt@rev.desc.d3.sspro.p2.T.catalog-pooled.boot-resid.value\endcsname{$+$0.9}
\expandafter\def\csname val@rev.desc.d3.sspro.p2.T.catalog-pooled.boot-share.ci95.0\endcsname{0.807}
\expandafter\def\csname cnt@rev.desc.d3.sspro.p2.T.catalog-pooled.boot-share.ci95.0\endcsname{1}
\expandafter\def\csname pct@rev.desc.d3.sspro.p2.T.catalog-pooled.boot-share.ci95.0\endcsname{80.7}
\expandafter\def\csname ppt@rev.desc.d3.sspro.p2.T.catalog-pooled.boot-share.ci95.0\endcsname{$+$80.7}
\expandafter\def\csname val@rev.desc.d3.sspro.p2.T.catalog-pooled.boot-share.ci95.1\endcsname{0.922}
\expandafter\def\csname cnt@rev.desc.d3.sspro.p2.T.catalog-pooled.boot-share.ci95.1\endcsname{1}
\expandafter\def\csname pct@rev.desc.d3.sspro.p2.T.catalog-pooled.boot-share.ci95.1\endcsname{92.2}
\expandafter\def\csname ppt@rev.desc.d3.sspro.p2.T.catalog-pooled.boot-share.ci95.1\endcsname{$+$92.2}
\expandafter\def\csname val@rev.desc.d3.sspro.p2.T.catalog-pooled.boot-share.n-apps\endcsname{12}
\expandafter\def\csname cnt@rev.desc.d3.sspro.p2.T.catalog-pooled.boot-share.n-apps\endcsname{12}
\expandafter\def\csname pct@rev.desc.d3.sspro.p2.T.catalog-pooled.boot-share.n-apps\endcsname{1200.0}
\expandafter\def\csname ppt@rev.desc.d3.sspro.p2.T.catalog-pooled.boot-share.n-apps\endcsname{$+$1200.0}
\expandafter\def\csname val@rev.desc.d3.sspro.p2.T.catalog-pooled.boot-share.n-boot-valid\endcsname{4000}
\expandafter\def\csname cnt@rev.desc.d3.sspro.p2.T.catalog-pooled.boot-share.n-boot-valid\endcsname{4{,}000}
\expandafter\def\csname pct@rev.desc.d3.sspro.p2.T.catalog-pooled.boot-share.n-boot-valid\endcsname{400000.0}
\expandafter\def\csname ppt@rev.desc.d3.sspro.p2.T.catalog-pooled.boot-share.n-boot-valid\endcsname{$+$400000.0}
\expandafter\def\csname val@rev.desc.d3.sspro.p2.T.catalog-pooled.boot-share.value\endcsname{0.870}
\expandafter\def\csname cnt@rev.desc.d3.sspro.p2.T.catalog-pooled.boot-share.value\endcsname{1}
\expandafter\def\csname pct@rev.desc.d3.sspro.p2.T.catalog-pooled.boot-share.value\endcsname{87.0}
\expandafter\def\csname ppt@rev.desc.d3.sspro.p2.T.catalog-pooled.boot-share.value\endcsname{$+$87.0}
\expandafter\def\csname val@rev.desc.d3.sspro.p2.T.catalog-pooled.gap\endcsname{0.089}
\expandafter\def\csname cnt@rev.desc.d3.sspro.p2.T.catalog-pooled.gap\endcsname{0}
\expandafter\def\csname pct@rev.desc.d3.sspro.p2.T.catalog-pooled.gap\endcsname{8.9}
\expandafter\def\csname ppt@rev.desc.d3.sspro.p2.T.catalog-pooled.gap\endcsname{$+$8.9}
\expandafter\def\csname val@rev.desc.d3.sspro.p2.T.catalog-pooled.n\endcsname{384}
\expandafter\def\csname cnt@rev.desc.d3.sspro.p2.T.catalog-pooled.n\endcsname{384}
\expandafter\def\csname pct@rev.desc.d3.sspro.p2.T.catalog-pooled.n\endcsname{38400.0}
\expandafter\def\csname ppt@rev.desc.d3.sspro.p2.T.catalog-pooled.n\endcsname{$+$38400.0}
\expandafter\def\csname val@rev.desc.d3.sspro.p2.T.catalog-pooled.reproduces-frozen\endcsname{1}
\expandafter\def\csname cnt@rev.desc.d3.sspro.p2.T.catalog-pooled.reproduces-frozen\endcsname{1}
\expandafter\def\csname pct@rev.desc.d3.sspro.p2.T.catalog-pooled.reproduces-frozen\endcsname{100.0}
\expandafter\def\csname ppt@rev.desc.d3.sspro.p2.T.catalog-pooled.reproduces-frozen\endcsname{$+$100.0}
\expandafter\def\csname val@rev.desc.d3.sspro.p2.T.catalog-pooled.resid-agree-stable\endcsname{0.009}
\expandafter\def\csname cnt@rev.desc.d3.sspro.p2.T.catalog-pooled.resid-agree-stable\endcsname{0}
\expandafter\def\csname pct@rev.desc.d3.sspro.p2.T.catalog-pooled.resid-agree-stable\endcsname{0.9}
\expandafter\def\csname ppt@rev.desc.d3.sspro.p2.T.catalog-pooled.resid-agree-stable\endcsname{$+$0.9}
\expandafter\def\csname val@rev.desc.d3.sspro.p2.T.catalog-pooled.resid-disagree\endcsname{0.098}
\expandafter\def\csname cnt@rev.desc.d3.sspro.p2.T.catalog-pooled.resid-disagree\endcsname{0}
\expandafter\def\csname pct@rev.desc.d3.sspro.p2.T.catalog-pooled.resid-disagree\endcsname{9.8}
\expandafter\def\csname ppt@rev.desc.d3.sspro.p2.T.catalog-pooled.resid-disagree\endcsname{$+$9.8}
\expandafter\def\csname val@rev.desc.d3.sspro.p2.T.catalog-pooled.share-agree-stable\endcsname{0.870}
\expandafter\def\csname cnt@rev.desc.d3.sspro.p2.T.catalog-pooled.share-agree-stable\endcsname{1}
\expandafter\def\csname pct@rev.desc.d3.sspro.p2.T.catalog-pooled.share-agree-stable\endcsname{87.0}
\expandafter\def\csname ppt@rev.desc.d3.sspro.p2.T.catalog-pooled.share-agree-stable\endcsname{$+$87.0}
\expandafter\def\csname val@rev.desc.d3.sspro.p2.T.center-d1.N-agree-stable\endcsname{87}
\expandafter\def\csname cnt@rev.desc.d3.sspro.p2.T.center-d1.N-agree-stable\endcsname{87}
\expandafter\def\csname pct@rev.desc.d3.sspro.p2.T.center-d1.N-agree-stable\endcsname{8700.0}
\expandafter\def\csname ppt@rev.desc.d3.sspro.p2.T.center-d1.N-agree-stable\endcsname{$+$8700.0}
\expandafter\def\csname val@rev.desc.d3.sspro.p2.T.center-d1.N-disagree\endcsname{8}
\expandafter\def\csname cnt@rev.desc.d3.sspro.p2.T.center-d1.N-disagree\endcsname{8}
\expandafter\def\csname pct@rev.desc.d3.sspro.p2.T.center-d1.N-disagree\endcsname{800.0}
\expandafter\def\csname ppt@rev.desc.d3.sspro.p2.T.center-d1.N-disagree\endcsname{$+$800.0}
\expandafter\def\csname val@rev.desc.d3.sspro.p2.T.center-d1.boot-resid.ci95.0\endcsname{-0.023}
\expandafter\def\csname cnt@rev.desc.d3.sspro.p2.T.center-d1.boot-resid.ci95.0\endcsname{0}
\expandafter\def\csname pct@rev.desc.d3.sspro.p2.T.center-d1.boot-resid.ci95.0\endcsname{$-$2.3}
\expandafter\def\csname ppt@rev.desc.d3.sspro.p2.T.center-d1.boot-resid.ci95.0\endcsname{$-$2.3}
\expandafter\def\csname val@rev.desc.d3.sspro.p2.T.center-d1.boot-resid.ci95.1\endcsname{0.048}
\expandafter\def\csname cnt@rev.desc.d3.sspro.p2.T.center-d1.boot-resid.ci95.1\endcsname{0}
\expandafter\def\csname pct@rev.desc.d3.sspro.p2.T.center-d1.boot-resid.ci95.1\endcsname{4.8}
\expandafter\def\csname ppt@rev.desc.d3.sspro.p2.T.center-d1.boot-resid.ci95.1\endcsname{$+$4.8}
\expandafter\def\csname val@rev.desc.d3.sspro.p2.T.center-d1.boot-resid.n-apps\endcsname{12}
\expandafter\def\csname cnt@rev.desc.d3.sspro.p2.T.center-d1.boot-resid.n-apps\endcsname{12}
\expandafter\def\csname pct@rev.desc.d3.sspro.p2.T.center-d1.boot-resid.n-apps\endcsname{1200.0}
\expandafter\def\csname ppt@rev.desc.d3.sspro.p2.T.center-d1.boot-resid.n-apps\endcsname{$+$1200.0}
\expandafter\def\csname val@rev.desc.d3.sspro.p2.T.center-d1.boot-resid.n-boot-valid\endcsname{4000}
\expandafter\def\csname cnt@rev.desc.d3.sspro.p2.T.center-d1.boot-resid.n-boot-valid\endcsname{4{,}000}
\expandafter\def\csname pct@rev.desc.d3.sspro.p2.T.center-d1.boot-resid.n-boot-valid\endcsname{400000.0}
\expandafter\def\csname ppt@rev.desc.d3.sspro.p2.T.center-d1.boot-resid.n-boot-valid\endcsname{$+$400000.0}
\expandafter\def\csname val@rev.desc.d3.sspro.p2.T.center-d1.boot-resid.value\endcsname{0.011}
\expandafter\def\csname cnt@rev.desc.d3.sspro.p2.T.center-d1.boot-resid.value\endcsname{0}
\expandafter\def\csname pct@rev.desc.d3.sspro.p2.T.center-d1.boot-resid.value\endcsname{1.1}
\expandafter\def\csname ppt@rev.desc.d3.sspro.p2.T.center-d1.boot-resid.value\endcsname{$+$1.1}
\expandafter\def\csname val@rev.desc.d3.sspro.p2.T.center-d1.boot-share.ci95.0\endcsname{0.844}
\expandafter\def\csname cnt@rev.desc.d3.sspro.p2.T.center-d1.boot-share.ci95.0\endcsname{1}
\expandafter\def\csname pct@rev.desc.d3.sspro.p2.T.center-d1.boot-share.ci95.0\endcsname{84.4}
\expandafter\def\csname ppt@rev.desc.d3.sspro.p2.T.center-d1.boot-share.ci95.0\endcsname{$+$84.4}
\expandafter\def\csname val@rev.desc.d3.sspro.p2.T.center-d1.boot-share.ci95.1\endcsname{0.958}
\expandafter\def\csname cnt@rev.desc.d3.sspro.p2.T.center-d1.boot-share.ci95.1\endcsname{1}
\expandafter\def\csname pct@rev.desc.d3.sspro.p2.T.center-d1.boot-share.ci95.1\endcsname{95.8}
\expandafter\def\csname ppt@rev.desc.d3.sspro.p2.T.center-d1.boot-share.ci95.1\endcsname{$+$95.8}
\expandafter\def\csname val@rev.desc.d3.sspro.p2.T.center-d1.boot-share.n-apps\endcsname{12}
\expandafter\def\csname cnt@rev.desc.d3.sspro.p2.T.center-d1.boot-share.n-apps\endcsname{12}
\expandafter\def\csname pct@rev.desc.d3.sspro.p2.T.center-d1.boot-share.n-apps\endcsname{1200.0}
\expandafter\def\csname ppt@rev.desc.d3.sspro.p2.T.center-d1.boot-share.n-apps\endcsname{$+$1200.0}
\expandafter\def\csname val@rev.desc.d3.sspro.p2.T.center-d1.boot-share.n-boot-valid\endcsname{4000}
\expandafter\def\csname cnt@rev.desc.d3.sspro.p2.T.center-d1.boot-share.n-boot-valid\endcsname{4{,}000}
\expandafter\def\csname pct@rev.desc.d3.sspro.p2.T.center-d1.boot-share.n-boot-valid\endcsname{400000.0}
\expandafter\def\csname ppt@rev.desc.d3.sspro.p2.T.center-d1.boot-share.n-boot-valid\endcsname{$+$400000.0}
\expandafter\def\csname val@rev.desc.d3.sspro.p2.T.center-d1.boot-share.value\endcsname{0.906}
\expandafter\def\csname cnt@rev.desc.d3.sspro.p2.T.center-d1.boot-share.value\endcsname{1}
\expandafter\def\csname pct@rev.desc.d3.sspro.p2.T.center-d1.boot-share.value\endcsname{90.6}
\expandafter\def\csname ppt@rev.desc.d3.sspro.p2.T.center-d1.boot-share.value\endcsname{$+$90.6}
\expandafter\def\csname val@rev.desc.d3.sspro.p2.T.center-d1.n\endcsname{96}
\expandafter\def\csname cnt@rev.desc.d3.sspro.p2.T.center-d1.n\endcsname{96}
\expandafter\def\csname pct@rev.desc.d3.sspro.p2.T.center-d1.n\endcsname{9600.0}
\expandafter\def\csname ppt@rev.desc.d3.sspro.p2.T.center-d1.n\endcsname{$+$9600.0}
\expandafter\def\csname val@rev.desc.d3.sspro.p2.T.center-d1.reproduces-frozen\endcsname{1}
\expandafter\def\csname cnt@rev.desc.d3.sspro.p2.T.center-d1.reproduces-frozen\endcsname{1}
\expandafter\def\csname pct@rev.desc.d3.sspro.p2.T.center-d1.reproduces-frozen\endcsname{100.0}
\expandafter\def\csname ppt@rev.desc.d3.sspro.p2.T.center-d1.reproduces-frozen\endcsname{$+$100.0}
\expandafter\def\csname val@rev.desc.d3.sspro.p2.T.center-d1.resid-agree-stable\endcsname{0.011}
\expandafter\def\csname cnt@rev.desc.d3.sspro.p2.T.center-d1.resid-agree-stable\endcsname{0}
\expandafter\def\csname pct@rev.desc.d3.sspro.p2.T.center-d1.resid-agree-stable\endcsname{1.1}
\expandafter\def\csname ppt@rev.desc.d3.sspro.p2.T.center-d1.resid-agree-stable\endcsname{$+$1.1}
\expandafter\def\csname val@rev.desc.d3.sspro.p2.T.center-d1.resid-disagree\endcsname{0.250}
\expandafter\def\csname cnt@rev.desc.d3.sspro.p2.T.center-d1.resid-disagree\endcsname{0}
\expandafter\def\csname pct@rev.desc.d3.sspro.p2.T.center-d1.resid-disagree\endcsname{25.0}
\expandafter\def\csname ppt@rev.desc.d3.sspro.p2.T.center-d1.resid-disagree\endcsname{$+$25.0}
\expandafter\def\csname val@rev.desc.d3.sspro.p2.T.center-d1.share-agree-stable\endcsname{0.906}
\expandafter\def\csname cnt@rev.desc.d3.sspro.p2.T.center-d1.share-agree-stable\endcsname{1}
\expandafter\def\csname pct@rev.desc.d3.sspro.p2.T.center-d1.share-agree-stable\endcsname{90.6}
\expandafter\def\csname ppt@rev.desc.d3.sspro.p2.T.center-d1.share-agree-stable\endcsname{$+$90.6}
\expandafter\def\csname val@rev.desc.d3.sspro.p2.T.context-d1.N-agree-stable\endcsname{86}
\expandafter\def\csname cnt@rev.desc.d3.sspro.p2.T.context-d1.N-agree-stable\endcsname{86}
\expandafter\def\csname pct@rev.desc.d3.sspro.p2.T.context-d1.N-agree-stable\endcsname{8600.0}
\expandafter\def\csname ppt@rev.desc.d3.sspro.p2.T.context-d1.N-agree-stable\endcsname{$+$8600.0}
\expandafter\def\csname val@rev.desc.d3.sspro.p2.T.context-d1.N-disagree\endcsname{8}
\expandafter\def\csname cnt@rev.desc.d3.sspro.p2.T.context-d1.N-disagree\endcsname{8}
\expandafter\def\csname pct@rev.desc.d3.sspro.p2.T.context-d1.N-disagree\endcsname{800.0}
\expandafter\def\csname ppt@rev.desc.d3.sspro.p2.T.context-d1.N-disagree\endcsname{$+$800.0}
\expandafter\def\csname val@rev.desc.d3.sspro.p2.T.context-d1.boot-resid.ci95.0\endcsname{-0.068}
\expandafter\def\csname cnt@rev.desc.d3.sspro.p2.T.context-d1.boot-resid.ci95.0\endcsname{0}
\expandafter\def\csname pct@rev.desc.d3.sspro.p2.T.context-d1.boot-resid.ci95.0\endcsname{$-$6.8}
\expandafter\def\csname ppt@rev.desc.d3.sspro.p2.T.context-d1.boot-resid.ci95.0\endcsname{$-$6.8}
\expandafter\def\csname val@rev.desc.d3.sspro.p2.T.context-d1.boot-resid.ci95.1\endcsname{0.022}
\expandafter\def\csname cnt@rev.desc.d3.sspro.p2.T.context-d1.boot-resid.ci95.1\endcsname{0}
\expandafter\def\csname pct@rev.desc.d3.sspro.p2.T.context-d1.boot-resid.ci95.1\endcsname{2.2}
\expandafter\def\csname ppt@rev.desc.d3.sspro.p2.T.context-d1.boot-resid.ci95.1\endcsname{$+$2.2}
\expandafter\def\csname val@rev.desc.d3.sspro.p2.T.context-d1.boot-resid.n-apps\endcsname{12}
\expandafter\def\csname cnt@rev.desc.d3.sspro.p2.T.context-d1.boot-resid.n-apps\endcsname{12}
\expandafter\def\csname pct@rev.desc.d3.sspro.p2.T.context-d1.boot-resid.n-apps\endcsname{1200.0}
\expandafter\def\csname ppt@rev.desc.d3.sspro.p2.T.context-d1.boot-resid.n-apps\endcsname{$+$1200.0}
\expandafter\def\csname val@rev.desc.d3.sspro.p2.T.context-d1.boot-resid.n-boot-valid\endcsname{4000}
\expandafter\def\csname cnt@rev.desc.d3.sspro.p2.T.context-d1.boot-resid.n-boot-valid\endcsname{4{,}000}
\expandafter\def\csname pct@rev.desc.d3.sspro.p2.T.context-d1.boot-resid.n-boot-valid\endcsname{400000.0}
\expandafter\def\csname ppt@rev.desc.d3.sspro.p2.T.context-d1.boot-resid.n-boot-valid\endcsname{$+$400000.0}
\expandafter\def\csname val@rev.desc.d3.sspro.p2.T.context-d1.boot-resid.value\endcsname{-0.023}
\expandafter\def\csname cnt@rev.desc.d3.sspro.p2.T.context-d1.boot-resid.value\endcsname{0}
\expandafter\def\csname pct@rev.desc.d3.sspro.p2.T.context-d1.boot-resid.value\endcsname{$-$2.3}
\expandafter\def\csname ppt@rev.desc.d3.sspro.p2.T.context-d1.boot-resid.value\endcsname{$-$2.3}
\expandafter\def\csname val@rev.desc.d3.sspro.p2.T.context-d1.boot-share.ci95.0\endcsname{0.833}
\expandafter\def\csname cnt@rev.desc.d3.sspro.p2.T.context-d1.boot-share.ci95.0\endcsname{1}
\expandafter\def\csname pct@rev.desc.d3.sspro.p2.T.context-d1.boot-share.ci95.0\endcsname{83.3}
\expandafter\def\csname ppt@rev.desc.d3.sspro.p2.T.context-d1.boot-share.ci95.0\endcsname{$+$83.3}
\expandafter\def\csname val@rev.desc.d3.sspro.p2.T.context-d1.boot-share.ci95.1\endcsname{0.948}
\expandafter\def\csname cnt@rev.desc.d3.sspro.p2.T.context-d1.boot-share.ci95.1\endcsname{1}
\expandafter\def\csname pct@rev.desc.d3.sspro.p2.T.context-d1.boot-share.ci95.1\endcsname{94.8}
\expandafter\def\csname ppt@rev.desc.d3.sspro.p2.T.context-d1.boot-share.ci95.1\endcsname{$+$94.8}
\expandafter\def\csname val@rev.desc.d3.sspro.p2.T.context-d1.boot-share.n-apps\endcsname{12}
\expandafter\def\csname cnt@rev.desc.d3.sspro.p2.T.context-d1.boot-share.n-apps\endcsname{12}
\expandafter\def\csname pct@rev.desc.d3.sspro.p2.T.context-d1.boot-share.n-apps\endcsname{1200.0}
\expandafter\def\csname ppt@rev.desc.d3.sspro.p2.T.context-d1.boot-share.n-apps\endcsname{$+$1200.0}
\expandafter\def\csname val@rev.desc.d3.sspro.p2.T.context-d1.boot-share.n-boot-valid\endcsname{4000}
\expandafter\def\csname cnt@rev.desc.d3.sspro.p2.T.context-d1.boot-share.n-boot-valid\endcsname{4{,}000}
\expandafter\def\csname pct@rev.desc.d3.sspro.p2.T.context-d1.boot-share.n-boot-valid\endcsname{400000.0}
\expandafter\def\csname ppt@rev.desc.d3.sspro.p2.T.context-d1.boot-share.n-boot-valid\endcsname{$+$400000.0}
\expandafter\def\csname val@rev.desc.d3.sspro.p2.T.context-d1.boot-share.value\endcsname{0.896}
\expandafter\def\csname cnt@rev.desc.d3.sspro.p2.T.context-d1.boot-share.value\endcsname{1}
\expandafter\def\csname pct@rev.desc.d3.sspro.p2.T.context-d1.boot-share.value\endcsname{89.6}
\expandafter\def\csname ppt@rev.desc.d3.sspro.p2.T.context-d1.boot-share.value\endcsname{$+$89.6}
\expandafter\def\csname val@rev.desc.d3.sspro.p2.T.context-d1.n\endcsname{96}
\expandafter\def\csname cnt@rev.desc.d3.sspro.p2.T.context-d1.n\endcsname{96}
\expandafter\def\csname pct@rev.desc.d3.sspro.p2.T.context-d1.n\endcsname{9600.0}
\expandafter\def\csname ppt@rev.desc.d3.sspro.p2.T.context-d1.n\endcsname{$+$9600.0}
\expandafter\def\csname val@rev.desc.d3.sspro.p2.T.context-d1.reproduces-frozen\endcsname{1}
\expandafter\def\csname cnt@rev.desc.d3.sspro.p2.T.context-d1.reproduces-frozen\endcsname{1}
\expandafter\def\csname pct@rev.desc.d3.sspro.p2.T.context-d1.reproduces-frozen\endcsname{100.0}
\expandafter\def\csname ppt@rev.desc.d3.sspro.p2.T.context-d1.reproduces-frozen\endcsname{$+$100.0}
\expandafter\def\csname val@rev.desc.d3.sspro.p2.T.context-d1.resid-agree-stable\endcsname{-0.023}
\expandafter\def\csname cnt@rev.desc.d3.sspro.p2.T.context-d1.resid-agree-stable\endcsname{0}
\expandafter\def\csname pct@rev.desc.d3.sspro.p2.T.context-d1.resid-agree-stable\endcsname{$-$2.3}
\expandafter\def\csname ppt@rev.desc.d3.sspro.p2.T.context-d1.resid-agree-stable\endcsname{$-$2.3}
\expandafter\def\csname val@rev.desc.d3.sspro.p2.T.context-d1.resid-disagree\endcsname{0.000}
\expandafter\def\csname cnt@rev.desc.d3.sspro.p2.T.context-d1.resid-disagree\endcsname{0}
\expandafter\def\csname pct@rev.desc.d3.sspro.p2.T.context-d1.resid-disagree\endcsname{0.0}
\expandafter\def\csname ppt@rev.desc.d3.sspro.p2.T.context-d1.resid-disagree\endcsname{0.0}
\expandafter\def\csname val@rev.desc.d3.sspro.p2.T.context-d1.share-agree-stable\endcsname{0.896}
\expandafter\def\csname cnt@rev.desc.d3.sspro.p2.T.context-d1.share-agree-stable\endcsname{1}
\expandafter\def\csname pct@rev.desc.d3.sspro.p2.T.context-d1.share-agree-stable\endcsname{89.6}
\expandafter\def\csname ppt@rev.desc.d3.sspro.p2.T.context-d1.share-agree-stable\endcsname{$+$89.6}
\expandafter\def\csname val@rev.desc.d3.sspro.p2.T.dual-d2.N-agree-stable\endcsname{82}
\expandafter\def\csname cnt@rev.desc.d3.sspro.p2.T.dual-d2.N-agree-stable\endcsname{82}
\expandafter\def\csname pct@rev.desc.d3.sspro.p2.T.dual-d2.N-agree-stable\endcsname{8200.0}
\expandafter\def\csname ppt@rev.desc.d3.sspro.p2.T.dual-d2.N-agree-stable\endcsname{$+$8200.0}
\expandafter\def\csname val@rev.desc.d3.sspro.p2.T.dual-d2.N-disagree\endcsname{12}
\expandafter\def\csname cnt@rev.desc.d3.sspro.p2.T.dual-d2.N-disagree\endcsname{12}
\expandafter\def\csname pct@rev.desc.d3.sspro.p2.T.dual-d2.N-disagree\endcsname{1200.0}
\expandafter\def\csname ppt@rev.desc.d3.sspro.p2.T.dual-d2.N-disagree\endcsname{$+$1200.0}
\expandafter\def\csname val@rev.desc.d3.sspro.p2.T.dual-d2.boot-resid.ci95.0\endcsname{0.000}
\expandafter\def\csname cnt@rev.desc.d3.sspro.p2.T.dual-d2.boot-resid.ci95.0\endcsname{0}
\expandafter\def\csname pct@rev.desc.d3.sspro.p2.T.dual-d2.boot-resid.ci95.0\endcsname{0.0}
\expandafter\def\csname ppt@rev.desc.d3.sspro.p2.T.dual-d2.boot-resid.ci95.0\endcsname{0.0}
\expandafter\def\csname val@rev.desc.d3.sspro.p2.T.dual-d2.boot-resid.ci95.1\endcsname{0.100}
\expandafter\def\csname cnt@rev.desc.d3.sspro.p2.T.dual-d2.boot-resid.ci95.1\endcsname{0}
\expandafter\def\csname pct@rev.desc.d3.sspro.p2.T.dual-d2.boot-resid.ci95.1\endcsname{10.0}
\expandafter\def\csname ppt@rev.desc.d3.sspro.p2.T.dual-d2.boot-resid.ci95.1\endcsname{$+$10.0}
\expandafter\def\csname val@rev.desc.d3.sspro.p2.T.dual-d2.boot-resid.n-apps\endcsname{12}
\expandafter\def\csname cnt@rev.desc.d3.sspro.p2.T.dual-d2.boot-resid.n-apps\endcsname{12}
\expandafter\def\csname pct@rev.desc.d3.sspro.p2.T.dual-d2.boot-resid.n-apps\endcsname{1200.0}
\expandafter\def\csname ppt@rev.desc.d3.sspro.p2.T.dual-d2.boot-resid.n-apps\endcsname{$+$1200.0}
\expandafter\def\csname val@rev.desc.d3.sspro.p2.T.dual-d2.boot-resid.n-boot-valid\endcsname{4000}
\expandafter\def\csname cnt@rev.desc.d3.sspro.p2.T.dual-d2.boot-resid.n-boot-valid\endcsname{4{,}000}
\expandafter\def\csname pct@rev.desc.d3.sspro.p2.T.dual-d2.boot-resid.n-boot-valid\endcsname{400000.0}
\expandafter\def\csname ppt@rev.desc.d3.sspro.p2.T.dual-d2.boot-resid.n-boot-valid\endcsname{$+$400000.0}
\expandafter\def\csname val@rev.desc.d3.sspro.p2.T.dual-d2.boot-resid.value\endcsname{0.049}
\expandafter\def\csname cnt@rev.desc.d3.sspro.p2.T.dual-d2.boot-resid.value\endcsname{0}
\expandafter\def\csname pct@rev.desc.d3.sspro.p2.T.dual-d2.boot-resid.value\endcsname{4.9}
\expandafter\def\csname ppt@rev.desc.d3.sspro.p2.T.dual-d2.boot-resid.value\endcsname{$+$4.9}
\expandafter\def\csname val@rev.desc.d3.sspro.p2.T.dual-d2.boot-share.ci95.0\endcsname{0.781}
\expandafter\def\csname cnt@rev.desc.d3.sspro.p2.T.dual-d2.boot-share.ci95.0\endcsname{1}
\expandafter\def\csname pct@rev.desc.d3.sspro.p2.T.dual-d2.boot-share.ci95.0\endcsname{78.1}
\expandafter\def\csname ppt@rev.desc.d3.sspro.p2.T.dual-d2.boot-share.ci95.0\endcsname{$+$78.1}
\expandafter\def\csname val@rev.desc.d3.sspro.p2.T.dual-d2.boot-share.ci95.1\endcsname{0.927}
\expandafter\def\csname cnt@rev.desc.d3.sspro.p2.T.dual-d2.boot-share.ci95.1\endcsname{1}
\expandafter\def\csname pct@rev.desc.d3.sspro.p2.T.dual-d2.boot-share.ci95.1\endcsname{92.7}
\expandafter\def\csname ppt@rev.desc.d3.sspro.p2.T.dual-d2.boot-share.ci95.1\endcsname{$+$92.7}
\expandafter\def\csname val@rev.desc.d3.sspro.p2.T.dual-d2.boot-share.n-apps\endcsname{12}
\expandafter\def\csname cnt@rev.desc.d3.sspro.p2.T.dual-d2.boot-share.n-apps\endcsname{12}
\expandafter\def\csname pct@rev.desc.d3.sspro.p2.T.dual-d2.boot-share.n-apps\endcsname{1200.0}
\expandafter\def\csname ppt@rev.desc.d3.sspro.p2.T.dual-d2.boot-share.n-apps\endcsname{$+$1200.0}
\expandafter\def\csname val@rev.desc.d3.sspro.p2.T.dual-d2.boot-share.n-boot-valid\endcsname{4000}
\expandafter\def\csname cnt@rev.desc.d3.sspro.p2.T.dual-d2.boot-share.n-boot-valid\endcsname{4{,}000}
\expandafter\def\csname pct@rev.desc.d3.sspro.p2.T.dual-d2.boot-share.n-boot-valid\endcsname{400000.0}
\expandafter\def\csname ppt@rev.desc.d3.sspro.p2.T.dual-d2.boot-share.n-boot-valid\endcsname{$+$400000.0}
\expandafter\def\csname val@rev.desc.d3.sspro.p2.T.dual-d2.boot-share.value\endcsname{0.854}
\expandafter\def\csname cnt@rev.desc.d3.sspro.p2.T.dual-d2.boot-share.value\endcsname{1}
\expandafter\def\csname pct@rev.desc.d3.sspro.p2.T.dual-d2.boot-share.value\endcsname{85.4}
\expandafter\def\csname ppt@rev.desc.d3.sspro.p2.T.dual-d2.boot-share.value\endcsname{$+$85.4}
\expandafter\def\csname val@rev.desc.d3.sspro.p2.T.dual-d2.n\endcsname{96}
\expandafter\def\csname cnt@rev.desc.d3.sspro.p2.T.dual-d2.n\endcsname{96}
\expandafter\def\csname pct@rev.desc.d3.sspro.p2.T.dual-d2.n\endcsname{9600.0}
\expandafter\def\csname ppt@rev.desc.d3.sspro.p2.T.dual-d2.n\endcsname{$+$9600.0}
\expandafter\def\csname val@rev.desc.d3.sspro.p2.T.dual-d2.reproduces-frozen\endcsname{1}
\expandafter\def\csname cnt@rev.desc.d3.sspro.p2.T.dual-d2.reproduces-frozen\endcsname{1}
\expandafter\def\csname pct@rev.desc.d3.sspro.p2.T.dual-d2.reproduces-frozen\endcsname{100.0}
\expandafter\def\csname ppt@rev.desc.d3.sspro.p2.T.dual-d2.reproduces-frozen\endcsname{$+$100.0}
\expandafter\def\csname val@rev.desc.d3.sspro.p2.T.dual-d2.resid-agree-stable\endcsname{0.049}
\expandafter\def\csname cnt@rev.desc.d3.sspro.p2.T.dual-d2.resid-agree-stable\endcsname{0}
\expandafter\def\csname pct@rev.desc.d3.sspro.p2.T.dual-d2.resid-agree-stable\endcsname{4.9}
\expandafter\def\csname ppt@rev.desc.d3.sspro.p2.T.dual-d2.resid-agree-stable\endcsname{$+$4.9}
\expandafter\def\csname val@rev.desc.d3.sspro.p2.T.dual-d2.resid-disagree\endcsname{0.083}
\expandafter\def\csname cnt@rev.desc.d3.sspro.p2.T.dual-d2.resid-disagree\endcsname{0}
\expandafter\def\csname pct@rev.desc.d3.sspro.p2.T.dual-d2.resid-disagree\endcsname{8.3}
\expandafter\def\csname ppt@rev.desc.d3.sspro.p2.T.dual-d2.resid-disagree\endcsname{$+$8.3}
\expandafter\def\csname val@rev.desc.d3.sspro.p2.T.dual-d2.share-agree-stable\endcsname{0.854}
\expandafter\def\csname cnt@rev.desc.d3.sspro.p2.T.dual-d2.share-agree-stable\endcsname{1}
\expandafter\def\csname pct@rev.desc.d3.sspro.p2.T.dual-d2.share-agree-stable\endcsname{85.4}
\expandafter\def\csname ppt@rev.desc.d3.sspro.p2.T.dual-d2.share-agree-stable\endcsname{$+$85.4}
\expandafter\def\csname val@rev.desc.d3.sspro.p2.T.grid2-d3.N-agree-stable\endcsname{79}
\expandafter\def\csname cnt@rev.desc.d3.sspro.p2.T.grid2-d3.N-agree-stable\endcsname{79}
\expandafter\def\csname pct@rev.desc.d3.sspro.p2.T.grid2-d3.N-agree-stable\endcsname{7900.0}
\expandafter\def\csname ppt@rev.desc.d3.sspro.p2.T.grid2-d3.N-agree-stable\endcsname{$+$7900.0}
\expandafter\def\csname val@rev.desc.d3.sspro.p2.T.grid2-d3.N-disagree\endcsname{13}
\expandafter\def\csname cnt@rev.desc.d3.sspro.p2.T.grid2-d3.N-disagree\endcsname{13}
\expandafter\def\csname pct@rev.desc.d3.sspro.p2.T.grid2-d3.N-disagree\endcsname{1300.0}
\expandafter\def\csname ppt@rev.desc.d3.sspro.p2.T.grid2-d3.N-disagree\endcsname{$+$1300.0}
\expandafter\def\csname val@rev.desc.d3.sspro.p2.T.grid2-d3.boot-resid.ci95.0\endcsname{-0.037}
\expandafter\def\csname cnt@rev.desc.d3.sspro.p2.T.grid2-d3.boot-resid.ci95.0\endcsname{0}
\expandafter\def\csname pct@rev.desc.d3.sspro.p2.T.grid2-d3.boot-resid.ci95.0\endcsname{$-$3.8}
\expandafter\def\csname ppt@rev.desc.d3.sspro.p2.T.grid2-d3.boot-resid.ci95.0\endcsname{$-$3.8}
\expandafter\def\csname val@rev.desc.d3.sspro.p2.T.grid2-d3.boot-resid.ci95.1\endcsname{0.037}
\expandafter\def\csname cnt@rev.desc.d3.sspro.p2.T.grid2-d3.boot-resid.ci95.1\endcsname{0}
\expandafter\def\csname pct@rev.desc.d3.sspro.p2.T.grid2-d3.boot-resid.ci95.1\endcsname{3.7}
\expandafter\def\csname ppt@rev.desc.d3.sspro.p2.T.grid2-d3.boot-resid.ci95.1\endcsname{$+$3.7}
\expandafter\def\csname val@rev.desc.d3.sspro.p2.T.grid2-d3.boot-resid.n-apps\endcsname{12}
\expandafter\def\csname cnt@rev.desc.d3.sspro.p2.T.grid2-d3.boot-resid.n-apps\endcsname{12}
\expandafter\def\csname pct@rev.desc.d3.sspro.p2.T.grid2-d3.boot-resid.n-apps\endcsname{1200.0}
\expandafter\def\csname ppt@rev.desc.d3.sspro.p2.T.grid2-d3.boot-resid.n-apps\endcsname{$+$1200.0}
\expandafter\def\csname val@rev.desc.d3.sspro.p2.T.grid2-d3.boot-resid.n-boot-valid\endcsname{4000}
\expandafter\def\csname cnt@rev.desc.d3.sspro.p2.T.grid2-d3.boot-resid.n-boot-valid\endcsname{4{,}000}
\expandafter\def\csname pct@rev.desc.d3.sspro.p2.T.grid2-d3.boot-resid.n-boot-valid\endcsname{400000.0}
\expandafter\def\csname ppt@rev.desc.d3.sspro.p2.T.grid2-d3.boot-resid.n-boot-valid\endcsname{$+$400000.0}
\expandafter\def\csname val@rev.desc.d3.sspro.p2.T.grid2-d3.boot-resid.value\endcsname{0.000}
\expandafter\def\csname cnt@rev.desc.d3.sspro.p2.T.grid2-d3.boot-resid.value\endcsname{0}
\expandafter\def\csname pct@rev.desc.d3.sspro.p2.T.grid2-d3.boot-resid.value\endcsname{0.0}
\expandafter\def\csname ppt@rev.desc.d3.sspro.p2.T.grid2-d3.boot-resid.value\endcsname{0.0}
\expandafter\def\csname val@rev.desc.d3.sspro.p2.T.grid2-d3.boot-share.ci95.0\endcsname{0.750}
\expandafter\def\csname cnt@rev.desc.d3.sspro.p2.T.grid2-d3.boot-share.ci95.0\endcsname{1}
\expandafter\def\csname pct@rev.desc.d3.sspro.p2.T.grid2-d3.boot-share.ci95.0\endcsname{75.0}
\expandafter\def\csname ppt@rev.desc.d3.sspro.p2.T.grid2-d3.boot-share.ci95.0\endcsname{$+$75.0}
\expandafter\def\csname val@rev.desc.d3.sspro.p2.T.grid2-d3.boot-share.ci95.1\endcsname{0.885}
\expandafter\def\csname cnt@rev.desc.d3.sspro.p2.T.grid2-d3.boot-share.ci95.1\endcsname{1}
\expandafter\def\csname pct@rev.desc.d3.sspro.p2.T.grid2-d3.boot-share.ci95.1\endcsname{88.5}
\expandafter\def\csname ppt@rev.desc.d3.sspro.p2.T.grid2-d3.boot-share.ci95.1\endcsname{$+$88.5}
\expandafter\def\csname val@rev.desc.d3.sspro.p2.T.grid2-d3.boot-share.n-apps\endcsname{12}
\expandafter\def\csname cnt@rev.desc.d3.sspro.p2.T.grid2-d3.boot-share.n-apps\endcsname{12}
\expandafter\def\csname pct@rev.desc.d3.sspro.p2.T.grid2-d3.boot-share.n-apps\endcsname{1200.0}
\expandafter\def\csname ppt@rev.desc.d3.sspro.p2.T.grid2-d3.boot-share.n-apps\endcsname{$+$1200.0}
\expandafter\def\csname val@rev.desc.d3.sspro.p2.T.grid2-d3.boot-share.n-boot-valid\endcsname{4000}
\expandafter\def\csname cnt@rev.desc.d3.sspro.p2.T.grid2-d3.boot-share.n-boot-valid\endcsname{4{,}000}
\expandafter\def\csname pct@rev.desc.d3.sspro.p2.T.grid2-d3.boot-share.n-boot-valid\endcsname{400000.0}
\expandafter\def\csname ppt@rev.desc.d3.sspro.p2.T.grid2-d3.boot-share.n-boot-valid\endcsname{$+$400000.0}
\expandafter\def\csname val@rev.desc.d3.sspro.p2.T.grid2-d3.boot-share.value\endcsname{0.823}
\expandafter\def\csname cnt@rev.desc.d3.sspro.p2.T.grid2-d3.boot-share.value\endcsname{1}
\expandafter\def\csname pct@rev.desc.d3.sspro.p2.T.grid2-d3.boot-share.value\endcsname{82.3}
\expandafter\def\csname ppt@rev.desc.d3.sspro.p2.T.grid2-d3.boot-share.value\endcsname{$+$82.3}
\expandafter\def\csname val@rev.desc.d3.sspro.p2.T.grid2-d3.n\endcsname{96}
\expandafter\def\csname cnt@rev.desc.d3.sspro.p2.T.grid2-d3.n\endcsname{96}
\expandafter\def\csname pct@rev.desc.d3.sspro.p2.T.grid2-d3.n\endcsname{9600.0}
\expandafter\def\csname ppt@rev.desc.d3.sspro.p2.T.grid2-d3.n\endcsname{$+$9600.0}
\expandafter\def\csname val@rev.desc.d3.sspro.p2.T.grid2-d3.reproduces-frozen\endcsname{1}
\expandafter\def\csname cnt@rev.desc.d3.sspro.p2.T.grid2-d3.reproduces-frozen\endcsname{1}
\expandafter\def\csname pct@rev.desc.d3.sspro.p2.T.grid2-d3.reproduces-frozen\endcsname{100.0}
\expandafter\def\csname ppt@rev.desc.d3.sspro.p2.T.grid2-d3.reproduces-frozen\endcsname{$+$100.0}
\expandafter\def\csname val@rev.desc.d3.sspro.p2.T.grid2-d3.resid-agree-stable\endcsname{0.000}
\expandafter\def\csname cnt@rev.desc.d3.sspro.p2.T.grid2-d3.resid-agree-stable\endcsname{0}
\expandafter\def\csname pct@rev.desc.d3.sspro.p2.T.grid2-d3.resid-agree-stable\endcsname{0.0}
\expandafter\def\csname ppt@rev.desc.d3.sspro.p2.T.grid2-d3.resid-agree-stable\endcsname{0.0}
\expandafter\def\csname val@rev.desc.d3.sspro.p2.T.grid2-d3.resid-disagree\endcsname{0.077}
\expandafter\def\csname cnt@rev.desc.d3.sspro.p2.T.grid2-d3.resid-disagree\endcsname{0}
\expandafter\def\csname pct@rev.desc.d3.sspro.p2.T.grid2-d3.resid-disagree\endcsname{7.7}
\expandafter\def\csname ppt@rev.desc.d3.sspro.p2.T.grid2-d3.resid-disagree\endcsname{$+$7.7}
\expandafter\def\csname val@rev.desc.d3.sspro.p2.T.grid2-d3.share-agree-stable\endcsname{0.823}
\expandafter\def\csname cnt@rev.desc.d3.sspro.p2.T.grid2-d3.share-agree-stable\endcsname{1}
\expandafter\def\csname pct@rev.desc.d3.sspro.p2.T.grid2-d3.share-agree-stable\endcsname{82.3}
\expandafter\def\csname ppt@rev.desc.d3.sspro.p2.T.grid2-d3.share-agree-stable\endcsname{$+$82.3}
\expandafter\def\csname val@rev.desc.d3.sspro.p2.T.zoomclick-d3.N-agree-stable\endcsname{82}
\expandafter\def\csname cnt@rev.desc.d3.sspro.p2.T.zoomclick-d3.N-agree-stable\endcsname{82}
\expandafter\def\csname pct@rev.desc.d3.sspro.p2.T.zoomclick-d3.N-agree-stable\endcsname{8200.0}
\expandafter\def\csname ppt@rev.desc.d3.sspro.p2.T.zoomclick-d3.N-agree-stable\endcsname{$+$8200.0}
\expandafter\def\csname val@rev.desc.d3.sspro.p2.T.zoomclick-d3.N-disagree\endcsname{12}
\expandafter\def\csname cnt@rev.desc.d3.sspro.p2.T.zoomclick-d3.N-disagree\endcsname{12}
\expandafter\def\csname pct@rev.desc.d3.sspro.p2.T.zoomclick-d3.N-disagree\endcsname{1200.0}
\expandafter\def\csname ppt@rev.desc.d3.sspro.p2.T.zoomclick-d3.N-disagree\endcsname{$+$1200.0}
\expandafter\def\csname val@rev.desc.d3.sspro.p2.T.zoomclick-d3.boot-resid.ci95.0\endcsname{0.000}
\expandafter\def\csname cnt@rev.desc.d3.sspro.p2.T.zoomclick-d3.boot-resid.ci95.0\endcsname{0}
\expandafter\def\csname pct@rev.desc.d3.sspro.p2.T.zoomclick-d3.boot-resid.ci95.0\endcsname{0.0}
\expandafter\def\csname ppt@rev.desc.d3.sspro.p2.T.zoomclick-d3.boot-resid.ci95.0\endcsname{0.0}
\expandafter\def\csname val@rev.desc.d3.sspro.p2.T.zoomclick-d3.boot-resid.ci95.1\endcsname{0.000}
\expandafter\def\csname cnt@rev.desc.d3.sspro.p2.T.zoomclick-d3.boot-resid.ci95.1\endcsname{0}
\expandafter\def\csname pct@rev.desc.d3.sspro.p2.T.zoomclick-d3.boot-resid.ci95.1\endcsname{0.0}
\expandafter\def\csname ppt@rev.desc.d3.sspro.p2.T.zoomclick-d3.boot-resid.ci95.1\endcsname{0.0}
\expandafter\def\csname val@rev.desc.d3.sspro.p2.T.zoomclick-d3.boot-resid.n-apps\endcsname{12}
\expandafter\def\csname cnt@rev.desc.d3.sspro.p2.T.zoomclick-d3.boot-resid.n-apps\endcsname{12}
\expandafter\def\csname pct@rev.desc.d3.sspro.p2.T.zoomclick-d3.boot-resid.n-apps\endcsname{1200.0}
\expandafter\def\csname ppt@rev.desc.d3.sspro.p2.T.zoomclick-d3.boot-resid.n-apps\endcsname{$+$1200.0}
\expandafter\def\csname val@rev.desc.d3.sspro.p2.T.zoomclick-d3.boot-resid.n-boot-valid\endcsname{4000}
\expandafter\def\csname cnt@rev.desc.d3.sspro.p2.T.zoomclick-d3.boot-resid.n-boot-valid\endcsname{4{,}000}
\expandafter\def\csname pct@rev.desc.d3.sspro.p2.T.zoomclick-d3.boot-resid.n-boot-valid\endcsname{400000.0}
\expandafter\def\csname ppt@rev.desc.d3.sspro.p2.T.zoomclick-d3.boot-resid.n-boot-valid\endcsname{$+$400000.0}
\expandafter\def\csname val@rev.desc.d3.sspro.p2.T.zoomclick-d3.boot-resid.value\endcsname{0.000}
\expandafter\def\csname cnt@rev.desc.d3.sspro.p2.T.zoomclick-d3.boot-resid.value\endcsname{0}
\expandafter\def\csname pct@rev.desc.d3.sspro.p2.T.zoomclick-d3.boot-resid.value\endcsname{0.0}
\expandafter\def\csname ppt@rev.desc.d3.sspro.p2.T.zoomclick-d3.boot-resid.value\endcsname{0.0}
\expandafter\def\csname val@rev.desc.d3.sspro.p2.T.zoomclick-d3.boot-share.ci95.0\endcsname{0.771}
\expandafter\def\csname cnt@rev.desc.d3.sspro.p2.T.zoomclick-d3.boot-share.ci95.0\endcsname{1}
\expandafter\def\csname pct@rev.desc.d3.sspro.p2.T.zoomclick-d3.boot-share.ci95.0\endcsname{77.1}
\expandafter\def\csname ppt@rev.desc.d3.sspro.p2.T.zoomclick-d3.boot-share.ci95.0\endcsname{$+$77.1}
\expandafter\def\csname val@rev.desc.d3.sspro.p2.T.zoomclick-d3.boot-share.ci95.1\endcsname{0.917}
\expandafter\def\csname cnt@rev.desc.d3.sspro.p2.T.zoomclick-d3.boot-share.ci95.1\endcsname{1}
\expandafter\def\csname pct@rev.desc.d3.sspro.p2.T.zoomclick-d3.boot-share.ci95.1\endcsname{91.7}
\expandafter\def\csname ppt@rev.desc.d3.sspro.p2.T.zoomclick-d3.boot-share.ci95.1\endcsname{$+$91.7}
\expandafter\def\csname val@rev.desc.d3.sspro.p2.T.zoomclick-d3.boot-share.n-apps\endcsname{12}
\expandafter\def\csname cnt@rev.desc.d3.sspro.p2.T.zoomclick-d3.boot-share.n-apps\endcsname{12}
\expandafter\def\csname pct@rev.desc.d3.sspro.p2.T.zoomclick-d3.boot-share.n-apps\endcsname{1200.0}
\expandafter\def\csname ppt@rev.desc.d3.sspro.p2.T.zoomclick-d3.boot-share.n-apps\endcsname{$+$1200.0}
\expandafter\def\csname val@rev.desc.d3.sspro.p2.T.zoomclick-d3.boot-share.n-boot-valid\endcsname{4000}
\expandafter\def\csname cnt@rev.desc.d3.sspro.p2.T.zoomclick-d3.boot-share.n-boot-valid\endcsname{4{,}000}
\expandafter\def\csname pct@rev.desc.d3.sspro.p2.T.zoomclick-d3.boot-share.n-boot-valid\endcsname{400000.0}
\expandafter\def\csname ppt@rev.desc.d3.sspro.p2.T.zoomclick-d3.boot-share.n-boot-valid\endcsname{$+$400000.0}
\expandafter\def\csname val@rev.desc.d3.sspro.p2.T.zoomclick-d3.boot-share.value\endcsname{0.854}
\expandafter\def\csname cnt@rev.desc.d3.sspro.p2.T.zoomclick-d3.boot-share.value\endcsname{1}
\expandafter\def\csname pct@rev.desc.d3.sspro.p2.T.zoomclick-d3.boot-share.value\endcsname{85.4}
\expandafter\def\csname ppt@rev.desc.d3.sspro.p2.T.zoomclick-d3.boot-share.value\endcsname{$+$85.4}
\expandafter\def\csname val@rev.desc.d3.sspro.p2.T.zoomclick-d3.n\endcsname{96}
\expandafter\def\csname cnt@rev.desc.d3.sspro.p2.T.zoomclick-d3.n\endcsname{96}
\expandafter\def\csname pct@rev.desc.d3.sspro.p2.T.zoomclick-d3.n\endcsname{9600.0}
\expandafter\def\csname ppt@rev.desc.d3.sspro.p2.T.zoomclick-d3.n\endcsname{$+$9600.0}
\expandafter\def\csname val@rev.desc.d3.sspro.p2.T.zoomclick-d3.reproduces-frozen\endcsname{1}
\expandafter\def\csname cnt@rev.desc.d3.sspro.p2.T.zoomclick-d3.reproduces-frozen\endcsname{1}
\expandafter\def\csname pct@rev.desc.d3.sspro.p2.T.zoomclick-d3.reproduces-frozen\endcsname{100.0}
\expandafter\def\csname ppt@rev.desc.d3.sspro.p2.T.zoomclick-d3.reproduces-frozen\endcsname{$+$100.0}
\expandafter\def\csname val@rev.desc.d3.sspro.p2.T.zoomclick-d3.resid-agree-stable\endcsname{0.000}
\expandafter\def\csname cnt@rev.desc.d3.sspro.p2.T.zoomclick-d3.resid-agree-stable\endcsname{0}
\expandafter\def\csname pct@rev.desc.d3.sspro.p2.T.zoomclick-d3.resid-agree-stable\endcsname{0.0}
\expandafter\def\csname ppt@rev.desc.d3.sspro.p2.T.zoomclick-d3.resid-agree-stable\endcsname{0.0}
\expandafter\def\csname val@rev.desc.d3.sspro.p2.T.zoomclick-d3.resid-disagree\endcsname{0.083}
\expandafter\def\csname cnt@rev.desc.d3.sspro.p2.T.zoomclick-d3.resid-disagree\endcsname{0}
\expandafter\def\csname pct@rev.desc.d3.sspro.p2.T.zoomclick-d3.resid-disagree\endcsname{8.3}
\expandafter\def\csname ppt@rev.desc.d3.sspro.p2.T.zoomclick-d3.resid-disagree\endcsname{$+$8.3}
\expandafter\def\csname val@rev.desc.d3.sspro.p2.T.zoomclick-d3.share-agree-stable\endcsname{0.854}
\expandafter\def\csname cnt@rev.desc.d3.sspro.p2.T.zoomclick-d3.share-agree-stable\endcsname{1}
\expandafter\def\csname pct@rev.desc.d3.sspro.p2.T.zoomclick-d3.share-agree-stable\endcsname{85.4}
\expandafter\def\csname ppt@rev.desc.d3.sspro.p2.T.zoomclick-d3.share-agree-stable\endcsname{$+$85.4}
\expandafter\def\csname val@rev.desc.d3.uiv.p1.C.catalog-pooled.N-agree-stable\endcsname{271}
\expandafter\def\csname cnt@rev.desc.d3.uiv.p1.C.catalog-pooled.N-agree-stable\endcsname{271}
\expandafter\def\csname pct@rev.desc.d3.uiv.p1.C.catalog-pooled.N-agree-stable\endcsname{27100.0}
\expandafter\def\csname ppt@rev.desc.d3.uiv.p1.C.catalog-pooled.N-agree-stable\endcsname{$+$27100.0}
\expandafter\def\csname val@rev.desc.d3.uiv.p1.C.catalog-pooled.N-disagree\endcsname{70}
\expandafter\def\csname cnt@rev.desc.d3.uiv.p1.C.catalog-pooled.N-disagree\endcsname{70}
\expandafter\def\csname pct@rev.desc.d3.uiv.p1.C.catalog-pooled.N-disagree\endcsname{7000.0}
\expandafter\def\csname ppt@rev.desc.d3.uiv.p1.C.catalog-pooled.N-disagree\endcsname{$+$7000.0}
\expandafter\def\csname val@rev.desc.d3.uiv.p1.C.catalog-pooled.boot-resid.ci95.0\endcsname{-0.004}
\expandafter\def\csname cnt@rev.desc.d3.uiv.p1.C.catalog-pooled.boot-resid.ci95.0\endcsname{0}
\expandafter\def\csname pct@rev.desc.d3.uiv.p1.C.catalog-pooled.boot-resid.ci95.0\endcsname{$-$0.4}
\expandafter\def\csname ppt@rev.desc.d3.uiv.p1.C.catalog-pooled.boot-resid.ci95.0\endcsname{$-$0.4}
\expandafter\def\csname val@rev.desc.d3.uiv.p1.C.catalog-pooled.boot-resid.ci95.1\endcsname{0.042}
\expandafter\def\csname cnt@rev.desc.d3.uiv.p1.C.catalog-pooled.boot-resid.ci95.1\endcsname{0}
\expandafter\def\csname pct@rev.desc.d3.uiv.p1.C.catalog-pooled.boot-resid.ci95.1\endcsname{4.2}
\expandafter\def\csname ppt@rev.desc.d3.uiv.p1.C.catalog-pooled.boot-resid.ci95.1\endcsname{$+$4.2}
\expandafter\def\csname val@rev.desc.d3.uiv.p1.C.catalog-pooled.boot-resid.n-apps\endcsname{12}
\expandafter\def\csname cnt@rev.desc.d3.uiv.p1.C.catalog-pooled.boot-resid.n-apps\endcsname{12}
\expandafter\def\csname pct@rev.desc.d3.uiv.p1.C.catalog-pooled.boot-resid.n-apps\endcsname{1200.0}
\expandafter\def\csname ppt@rev.desc.d3.uiv.p1.C.catalog-pooled.boot-resid.n-apps\endcsname{$+$1200.0}
\expandafter\def\csname val@rev.desc.d3.uiv.p1.C.catalog-pooled.boot-resid.n-boot-valid\endcsname{4000}
\expandafter\def\csname cnt@rev.desc.d3.uiv.p1.C.catalog-pooled.boot-resid.n-boot-valid\endcsname{4{,}000}
\expandafter\def\csname pct@rev.desc.d3.uiv.p1.C.catalog-pooled.boot-resid.n-boot-valid\endcsname{400000.0}
\expandafter\def\csname ppt@rev.desc.d3.uiv.p1.C.catalog-pooled.boot-resid.n-boot-valid\endcsname{$+$400000.0}
\expandafter\def\csname val@rev.desc.d3.uiv.p1.C.catalog-pooled.boot-resid.value\endcsname{0.018}
\expandafter\def\csname cnt@rev.desc.d3.uiv.p1.C.catalog-pooled.boot-resid.value\endcsname{0}
\expandafter\def\csname pct@rev.desc.d3.uiv.p1.C.catalog-pooled.boot-resid.value\endcsname{1.8}
\expandafter\def\csname ppt@rev.desc.d3.uiv.p1.C.catalog-pooled.boot-resid.value\endcsname{$+$1.8}
\expandafter\def\csname val@rev.desc.d3.uiv.p1.C.catalog-pooled.boot-share.ci95.0\endcsname{0.641}
\expandafter\def\csname cnt@rev.desc.d3.uiv.p1.C.catalog-pooled.boot-share.ci95.0\endcsname{1}
\expandafter\def\csname pct@rev.desc.d3.uiv.p1.C.catalog-pooled.boot-share.ci95.0\endcsname{64.1}
\expandafter\def\csname ppt@rev.desc.d3.uiv.p1.C.catalog-pooled.boot-share.ci95.0\endcsname{$+$64.1}
\expandafter\def\csname val@rev.desc.d3.uiv.p1.C.catalog-pooled.boot-share.ci95.1\endcsname{0.776}
\expandafter\def\csname cnt@rev.desc.d3.uiv.p1.C.catalog-pooled.boot-share.ci95.1\endcsname{1}
\expandafter\def\csname pct@rev.desc.d3.uiv.p1.C.catalog-pooled.boot-share.ci95.1\endcsname{77.6}
\expandafter\def\csname ppt@rev.desc.d3.uiv.p1.C.catalog-pooled.boot-share.ci95.1\endcsname{$+$77.6}
\expandafter\def\csname val@rev.desc.d3.uiv.p1.C.catalog-pooled.boot-share.n-apps\endcsname{12}
\expandafter\def\csname cnt@rev.desc.d3.uiv.p1.C.catalog-pooled.boot-share.n-apps\endcsname{12}
\expandafter\def\csname pct@rev.desc.d3.uiv.p1.C.catalog-pooled.boot-share.n-apps\endcsname{1200.0}
\expandafter\def\csname ppt@rev.desc.d3.uiv.p1.C.catalog-pooled.boot-share.n-apps\endcsname{$+$1200.0}
\expandafter\def\csname val@rev.desc.d3.uiv.p1.C.catalog-pooled.boot-share.n-boot-valid\endcsname{4000}
\expandafter\def\csname cnt@rev.desc.d3.uiv.p1.C.catalog-pooled.boot-share.n-boot-valid\endcsname{4{,}000}
\expandafter\def\csname pct@rev.desc.d3.uiv.p1.C.catalog-pooled.boot-share.n-boot-valid\endcsname{400000.0}
\expandafter\def\csname ppt@rev.desc.d3.uiv.p1.C.catalog-pooled.boot-share.n-boot-valid\endcsname{$+$400000.0}
\expandafter\def\csname val@rev.desc.d3.uiv.p1.C.catalog-pooled.boot-share.value\endcsname{0.706}
\expandafter\def\csname cnt@rev.desc.d3.uiv.p1.C.catalog-pooled.boot-share.value\endcsname{1}
\expandafter\def\csname pct@rev.desc.d3.uiv.p1.C.catalog-pooled.boot-share.value\endcsname{70.6}
\expandafter\def\csname ppt@rev.desc.d3.uiv.p1.C.catalog-pooled.boot-share.value\endcsname{$+$70.6}
\expandafter\def\csname val@rev.desc.d3.uiv.p1.C.catalog-pooled.gap\endcsname{0.253}
\expandafter\def\csname cnt@rev.desc.d3.uiv.p1.C.catalog-pooled.gap\endcsname{0}
\expandafter\def\csname pct@rev.desc.d3.uiv.p1.C.catalog-pooled.gap\endcsname{25.3}
\expandafter\def\csname ppt@rev.desc.d3.uiv.p1.C.catalog-pooled.gap\endcsname{$+$25.3}
\expandafter\def\csname val@rev.desc.d3.uiv.p1.C.catalog-pooled.n\endcsname{384}
\expandafter\def\csname cnt@rev.desc.d3.uiv.p1.C.catalog-pooled.n\endcsname{384}
\expandafter\def\csname pct@rev.desc.d3.uiv.p1.C.catalog-pooled.n\endcsname{38400.0}
\expandafter\def\csname ppt@rev.desc.d3.uiv.p1.C.catalog-pooled.n\endcsname{$+$38400.0}
\expandafter\def\csname val@rev.desc.d3.uiv.p1.C.catalog-pooled.reproduces-frozen\endcsname{1}
\expandafter\def\csname cnt@rev.desc.d3.uiv.p1.C.catalog-pooled.reproduces-frozen\endcsname{1}
\expandafter\def\csname pct@rev.desc.d3.uiv.p1.C.catalog-pooled.reproduces-frozen\endcsname{100.0}
\expandafter\def\csname ppt@rev.desc.d3.uiv.p1.C.catalog-pooled.reproduces-frozen\endcsname{$+$100.0}
\expandafter\def\csname val@rev.desc.d3.uiv.p1.C.catalog-pooled.resid-agree-stable\endcsname{0.018}
\expandafter\def\csname cnt@rev.desc.d3.uiv.p1.C.catalog-pooled.resid-agree-stable\endcsname{0}
\expandafter\def\csname pct@rev.desc.d3.uiv.p1.C.catalog-pooled.resid-agree-stable\endcsname{1.8}
\expandafter\def\csname ppt@rev.desc.d3.uiv.p1.C.catalog-pooled.resid-agree-stable\endcsname{$+$1.8}
\expandafter\def\csname val@rev.desc.d3.uiv.p1.C.catalog-pooled.resid-disagree\endcsname{0.271}
\expandafter\def\csname cnt@rev.desc.d3.uiv.p1.C.catalog-pooled.resid-disagree\endcsname{0}
\expandafter\def\csname pct@rev.desc.d3.uiv.p1.C.catalog-pooled.resid-disagree\endcsname{27.1}
\expandafter\def\csname ppt@rev.desc.d3.uiv.p1.C.catalog-pooled.resid-disagree\endcsname{$+$27.1}
\expandafter\def\csname val@rev.desc.d3.uiv.p1.C.catalog-pooled.share-agree-stable\endcsname{0.706}
\expandafter\def\csname cnt@rev.desc.d3.uiv.p1.C.catalog-pooled.share-agree-stable\endcsname{1}
\expandafter\def\csname pct@rev.desc.d3.uiv.p1.C.catalog-pooled.share-agree-stable\endcsname{70.6}
\expandafter\def\csname ppt@rev.desc.d3.uiv.p1.C.catalog-pooled.share-agree-stable\endcsname{$+$70.6}
\expandafter\def\csname val@rev.desc.d3.uiv.p1.C.center-d1.N-agree-stable\endcsname{70}
\expandafter\def\csname cnt@rev.desc.d3.uiv.p1.C.center-d1.N-agree-stable\endcsname{70}
\expandafter\def\csname pct@rev.desc.d3.uiv.p1.C.center-d1.N-agree-stable\endcsname{7000.0}
\expandafter\def\csname ppt@rev.desc.d3.uiv.p1.C.center-d1.N-agree-stable\endcsname{$+$7000.0}
\expandafter\def\csname val@rev.desc.d3.uiv.p1.C.center-d1.N-disagree\endcsname{17}
\expandafter\def\csname cnt@rev.desc.d3.uiv.p1.C.center-d1.N-disagree\endcsname{17}
\expandafter\def\csname pct@rev.desc.d3.uiv.p1.C.center-d1.N-disagree\endcsname{1700.0}
\expandafter\def\csname ppt@rev.desc.d3.uiv.p1.C.center-d1.N-disagree\endcsname{$+$1700.0}
\expandafter\def\csname val@rev.desc.d3.uiv.p1.C.center-d1.boot-resid.ci95.0\endcsname{0.000}
\expandafter\def\csname cnt@rev.desc.d3.uiv.p1.C.center-d1.boot-resid.ci95.0\endcsname{0}
\expandafter\def\csname pct@rev.desc.d3.uiv.p1.C.center-d1.boot-resid.ci95.0\endcsname{0.0}
\expandafter\def\csname ppt@rev.desc.d3.uiv.p1.C.center-d1.boot-resid.ci95.0\endcsname{0.0}
\expandafter\def\csname val@rev.desc.d3.uiv.p1.C.center-d1.boot-resid.ci95.1\endcsname{0.042}
\expandafter\def\csname cnt@rev.desc.d3.uiv.p1.C.center-d1.boot-resid.ci95.1\endcsname{0}
\expandafter\def\csname pct@rev.desc.d3.uiv.p1.C.center-d1.boot-resid.ci95.1\endcsname{4.2}
\expandafter\def\csname ppt@rev.desc.d3.uiv.p1.C.center-d1.boot-resid.ci95.1\endcsname{$+$4.2}
\expandafter\def\csname val@rev.desc.d3.uiv.p1.C.center-d1.boot-resid.n-apps\endcsname{12}
\expandafter\def\csname cnt@rev.desc.d3.uiv.p1.C.center-d1.boot-resid.n-apps\endcsname{12}
\expandafter\def\csname pct@rev.desc.d3.uiv.p1.C.center-d1.boot-resid.n-apps\endcsname{1200.0}
\expandafter\def\csname ppt@rev.desc.d3.uiv.p1.C.center-d1.boot-resid.n-apps\endcsname{$+$1200.0}
\expandafter\def\csname val@rev.desc.d3.uiv.p1.C.center-d1.boot-resid.n-boot-valid\endcsname{4000}
\expandafter\def\csname cnt@rev.desc.d3.uiv.p1.C.center-d1.boot-resid.n-boot-valid\endcsname{4{,}000}
\expandafter\def\csname pct@rev.desc.d3.uiv.p1.C.center-d1.boot-resid.n-boot-valid\endcsname{400000.0}
\expandafter\def\csname ppt@rev.desc.d3.uiv.p1.C.center-d1.boot-resid.n-boot-valid\endcsname{$+$400000.0}
\expandafter\def\csname val@rev.desc.d3.uiv.p1.C.center-d1.boot-resid.value\endcsname{0.014}
\expandafter\def\csname cnt@rev.desc.d3.uiv.p1.C.center-d1.boot-resid.value\endcsname{0}
\expandafter\def\csname pct@rev.desc.d3.uiv.p1.C.center-d1.boot-resid.value\endcsname{1.4}
\expandafter\def\csname ppt@rev.desc.d3.uiv.p1.C.center-d1.boot-resid.value\endcsname{$+$1.4}
\expandafter\def\csname val@rev.desc.d3.uiv.p1.C.center-d1.boot-share.ci95.0\endcsname{0.656}
\expandafter\def\csname cnt@rev.desc.d3.uiv.p1.C.center-d1.boot-share.ci95.0\endcsname{1}
\expandafter\def\csname pct@rev.desc.d3.uiv.p1.C.center-d1.boot-share.ci95.0\endcsname{65.6}
\expandafter\def\csname ppt@rev.desc.d3.uiv.p1.C.center-d1.boot-share.ci95.0\endcsname{$+$65.6}
\expandafter\def\csname val@rev.desc.d3.uiv.p1.C.center-d1.boot-share.ci95.1\endcsname{0.802}
\expandafter\def\csname cnt@rev.desc.d3.uiv.p1.C.center-d1.boot-share.ci95.1\endcsname{1}
\expandafter\def\csname pct@rev.desc.d3.uiv.p1.C.center-d1.boot-share.ci95.1\endcsname{80.2}
\expandafter\def\csname ppt@rev.desc.d3.uiv.p1.C.center-d1.boot-share.ci95.1\endcsname{$+$80.2}
\expandafter\def\csname val@rev.desc.d3.uiv.p1.C.center-d1.boot-share.n-apps\endcsname{12}
\expandafter\def\csname cnt@rev.desc.d3.uiv.p1.C.center-d1.boot-share.n-apps\endcsname{12}
\expandafter\def\csname pct@rev.desc.d3.uiv.p1.C.center-d1.boot-share.n-apps\endcsname{1200.0}
\expandafter\def\csname ppt@rev.desc.d3.uiv.p1.C.center-d1.boot-share.n-apps\endcsname{$+$1200.0}
\expandafter\def\csname val@rev.desc.d3.uiv.p1.C.center-d1.boot-share.n-boot-valid\endcsname{4000}
\expandafter\def\csname cnt@rev.desc.d3.uiv.p1.C.center-d1.boot-share.n-boot-valid\endcsname{4{,}000}
\expandafter\def\csname pct@rev.desc.d3.uiv.p1.C.center-d1.boot-share.n-boot-valid\endcsname{400000.0}
\expandafter\def\csname ppt@rev.desc.d3.uiv.p1.C.center-d1.boot-share.n-boot-valid\endcsname{$+$400000.0}
\expandafter\def\csname val@rev.desc.d3.uiv.p1.C.center-d1.boot-share.value\endcsname{0.729}
\expandafter\def\csname cnt@rev.desc.d3.uiv.p1.C.center-d1.boot-share.value\endcsname{1}
\expandafter\def\csname pct@rev.desc.d3.uiv.p1.C.center-d1.boot-share.value\endcsname{72.9}
\expandafter\def\csname ppt@rev.desc.d3.uiv.p1.C.center-d1.boot-share.value\endcsname{$+$72.9}
\expandafter\def\csname val@rev.desc.d3.uiv.p1.C.center-d1.n\endcsname{96}
\expandafter\def\csname cnt@rev.desc.d3.uiv.p1.C.center-d1.n\endcsname{96}
\expandafter\def\csname pct@rev.desc.d3.uiv.p1.C.center-d1.n\endcsname{9600.0}
\expandafter\def\csname ppt@rev.desc.d3.uiv.p1.C.center-d1.n\endcsname{$+$9600.0}
\expandafter\def\csname val@rev.desc.d3.uiv.p1.C.center-d1.reproduces-frozen\endcsname{1}
\expandafter\def\csname cnt@rev.desc.d3.uiv.p1.C.center-d1.reproduces-frozen\endcsname{1}
\expandafter\def\csname pct@rev.desc.d3.uiv.p1.C.center-d1.reproduces-frozen\endcsname{100.0}
\expandafter\def\csname ppt@rev.desc.d3.uiv.p1.C.center-d1.reproduces-frozen\endcsname{$+$100.0}
\expandafter\def\csname val@rev.desc.d3.uiv.p1.C.center-d1.resid-agree-stable\endcsname{0.014}
\expandafter\def\csname cnt@rev.desc.d3.uiv.p1.C.center-d1.resid-agree-stable\endcsname{0}
\expandafter\def\csname pct@rev.desc.d3.uiv.p1.C.center-d1.resid-agree-stable\endcsname{1.4}
\expandafter\def\csname ppt@rev.desc.d3.uiv.p1.C.center-d1.resid-agree-stable\endcsname{$+$1.4}
\expandafter\def\csname val@rev.desc.d3.uiv.p1.C.center-d1.resid-disagree\endcsname{0.353}
\expandafter\def\csname cnt@rev.desc.d3.uiv.p1.C.center-d1.resid-disagree\endcsname{0}
\expandafter\def\csname pct@rev.desc.d3.uiv.p1.C.center-d1.resid-disagree\endcsname{35.3}
\expandafter\def\csname ppt@rev.desc.d3.uiv.p1.C.center-d1.resid-disagree\endcsname{$+$35.3}
\expandafter\def\csname val@rev.desc.d3.uiv.p1.C.center-d1.share-agree-stable\endcsname{0.729}
\expandafter\def\csname cnt@rev.desc.d3.uiv.p1.C.center-d1.share-agree-stable\endcsname{1}
\expandafter\def\csname pct@rev.desc.d3.uiv.p1.C.center-d1.share-agree-stable\endcsname{72.9}
\expandafter\def\csname ppt@rev.desc.d3.uiv.p1.C.center-d1.share-agree-stable\endcsname{$+$72.9}
\expandafter\def\csname val@rev.desc.d3.uiv.p1.C.context-d1.N-agree-stable\endcsname{73}
\expandafter\def\csname cnt@rev.desc.d3.uiv.p1.C.context-d1.N-agree-stable\endcsname{73}
\expandafter\def\csname pct@rev.desc.d3.uiv.p1.C.context-d1.N-agree-stable\endcsname{7300.0}
\expandafter\def\csname ppt@rev.desc.d3.uiv.p1.C.context-d1.N-agree-stable\endcsname{$+$7300.0}
\expandafter\def\csname val@rev.desc.d3.uiv.p1.C.context-d1.N-disagree\endcsname{11}
\expandafter\def\csname cnt@rev.desc.d3.uiv.p1.C.context-d1.N-disagree\endcsname{11}
\expandafter\def\csname pct@rev.desc.d3.uiv.p1.C.context-d1.N-disagree\endcsname{1100.0}
\expandafter\def\csname ppt@rev.desc.d3.uiv.p1.C.context-d1.N-disagree\endcsname{$+$1100.0}
\expandafter\def\csname val@rev.desc.d3.uiv.p1.C.context-d1.boot-resid.ci95.0\endcsname{0.000}
\expandafter\def\csname cnt@rev.desc.d3.uiv.p1.C.context-d1.boot-resid.ci95.0\endcsname{0}
\expandafter\def\csname pct@rev.desc.d3.uiv.p1.C.context-d1.boot-resid.ci95.0\endcsname{0.0}
\expandafter\def\csname ppt@rev.desc.d3.uiv.p1.C.context-d1.boot-resid.ci95.0\endcsname{0.0}
\expandafter\def\csname val@rev.desc.d3.uiv.p1.C.context-d1.boot-resid.ci95.1\endcsname{0.041}
\expandafter\def\csname cnt@rev.desc.d3.uiv.p1.C.context-d1.boot-resid.ci95.1\endcsname{0}
\expandafter\def\csname pct@rev.desc.d3.uiv.p1.C.context-d1.boot-resid.ci95.1\endcsname{4.1}
\expandafter\def\csname ppt@rev.desc.d3.uiv.p1.C.context-d1.boot-resid.ci95.1\endcsname{$+$4.1}
\expandafter\def\csname val@rev.desc.d3.uiv.p1.C.context-d1.boot-resid.n-apps\endcsname{12}
\expandafter\def\csname cnt@rev.desc.d3.uiv.p1.C.context-d1.boot-resid.n-apps\endcsname{12}
\expandafter\def\csname pct@rev.desc.d3.uiv.p1.C.context-d1.boot-resid.n-apps\endcsname{1200.0}
\expandafter\def\csname ppt@rev.desc.d3.uiv.p1.C.context-d1.boot-resid.n-apps\endcsname{$+$1200.0}
\expandafter\def\csname val@rev.desc.d3.uiv.p1.C.context-d1.boot-resid.n-boot-valid\endcsname{4000}
\expandafter\def\csname cnt@rev.desc.d3.uiv.p1.C.context-d1.boot-resid.n-boot-valid\endcsname{4{,}000}
\expandafter\def\csname pct@rev.desc.d3.uiv.p1.C.context-d1.boot-resid.n-boot-valid\endcsname{400000.0}
\expandafter\def\csname ppt@rev.desc.d3.uiv.p1.C.context-d1.boot-resid.n-boot-valid\endcsname{$+$400000.0}
\expandafter\def\csname val@rev.desc.d3.uiv.p1.C.context-d1.boot-resid.value\endcsname{0.014}
\expandafter\def\csname cnt@rev.desc.d3.uiv.p1.C.context-d1.boot-resid.value\endcsname{0}
\expandafter\def\csname pct@rev.desc.d3.uiv.p1.C.context-d1.boot-resid.value\endcsname{1.4}
\expandafter\def\csname ppt@rev.desc.d3.uiv.p1.C.context-d1.boot-resid.value\endcsname{$+$1.4}
\expandafter\def\csname val@rev.desc.d3.uiv.p1.C.context-d1.boot-share.ci95.0\endcsname{0.677}
\expandafter\def\csname cnt@rev.desc.d3.uiv.p1.C.context-d1.boot-share.ci95.0\endcsname{1}
\expandafter\def\csname pct@rev.desc.d3.uiv.p1.C.context-d1.boot-share.ci95.0\endcsname{67.7}
\expandafter\def\csname ppt@rev.desc.d3.uiv.p1.C.context-d1.boot-share.ci95.0\endcsname{$+$67.7}
\expandafter\def\csname val@rev.desc.d3.uiv.p1.C.context-d1.boot-share.ci95.1\endcsname{0.844}
\expandafter\def\csname cnt@rev.desc.d3.uiv.p1.C.context-d1.boot-share.ci95.1\endcsname{1}
\expandafter\def\csname pct@rev.desc.d3.uiv.p1.C.context-d1.boot-share.ci95.1\endcsname{84.4}
\expandafter\def\csname ppt@rev.desc.d3.uiv.p1.C.context-d1.boot-share.ci95.1\endcsname{$+$84.4}
\expandafter\def\csname val@rev.desc.d3.uiv.p1.C.context-d1.boot-share.n-apps\endcsname{12}
\expandafter\def\csname cnt@rev.desc.d3.uiv.p1.C.context-d1.boot-share.n-apps\endcsname{12}
\expandafter\def\csname pct@rev.desc.d3.uiv.p1.C.context-d1.boot-share.n-apps\endcsname{1200.0}
\expandafter\def\csname ppt@rev.desc.d3.uiv.p1.C.context-d1.boot-share.n-apps\endcsname{$+$1200.0}
\expandafter\def\csname val@rev.desc.d3.uiv.p1.C.context-d1.boot-share.n-boot-valid\endcsname{4000}
\expandafter\def\csname cnt@rev.desc.d3.uiv.p1.C.context-d1.boot-share.n-boot-valid\endcsname{4{,}000}
\expandafter\def\csname pct@rev.desc.d3.uiv.p1.C.context-d1.boot-share.n-boot-valid\endcsname{400000.0}
\expandafter\def\csname ppt@rev.desc.d3.uiv.p1.C.context-d1.boot-share.n-boot-valid\endcsname{$+$400000.0}
\expandafter\def\csname val@rev.desc.d3.uiv.p1.C.context-d1.boot-share.value\endcsname{0.760}
\expandafter\def\csname cnt@rev.desc.d3.uiv.p1.C.context-d1.boot-share.value\endcsname{1}
\expandafter\def\csname pct@rev.desc.d3.uiv.p1.C.context-d1.boot-share.value\endcsname{76.0}
\expandafter\def\csname ppt@rev.desc.d3.uiv.p1.C.context-d1.boot-share.value\endcsname{$+$76.0}
\expandafter\def\csname val@rev.desc.d3.uiv.p1.C.context-d1.n\endcsname{96}
\expandafter\def\csname cnt@rev.desc.d3.uiv.p1.C.context-d1.n\endcsname{96}
\expandafter\def\csname pct@rev.desc.d3.uiv.p1.C.context-d1.n\endcsname{9600.0}
\expandafter\def\csname ppt@rev.desc.d3.uiv.p1.C.context-d1.n\endcsname{$+$9600.0}
\expandafter\def\csname val@rev.desc.d3.uiv.p1.C.context-d1.reproduces-frozen\endcsname{1}
\expandafter\def\csname cnt@rev.desc.d3.uiv.p1.C.context-d1.reproduces-frozen\endcsname{1}
\expandafter\def\csname pct@rev.desc.d3.uiv.p1.C.context-d1.reproduces-frozen\endcsname{100.0}
\expandafter\def\csname ppt@rev.desc.d3.uiv.p1.C.context-d1.reproduces-frozen\endcsname{$+$100.0}
\expandafter\def\csname val@rev.desc.d3.uiv.p1.C.context-d1.resid-agree-stable\endcsname{0.014}
\expandafter\def\csname cnt@rev.desc.d3.uiv.p1.C.context-d1.resid-agree-stable\endcsname{0}
\expandafter\def\csname pct@rev.desc.d3.uiv.p1.C.context-d1.resid-agree-stable\endcsname{1.4}
\expandafter\def\csname ppt@rev.desc.d3.uiv.p1.C.context-d1.resid-agree-stable\endcsname{$+$1.4}
\expandafter\def\csname val@rev.desc.d3.uiv.p1.C.context-d1.resid-disagree\endcsname{-0.091}
\expandafter\def\csname cnt@rev.desc.d3.uiv.p1.C.context-d1.resid-disagree\endcsname{0}
\expandafter\def\csname pct@rev.desc.d3.uiv.p1.C.context-d1.resid-disagree\endcsname{$-$9.1}
\expandafter\def\csname ppt@rev.desc.d3.uiv.p1.C.context-d1.resid-disagree\endcsname{$-$9.1}
\expandafter\def\csname val@rev.desc.d3.uiv.p1.C.context-d1.share-agree-stable\endcsname{0.760}
\expandafter\def\csname cnt@rev.desc.d3.uiv.p1.C.context-d1.share-agree-stable\endcsname{1}
\expandafter\def\csname pct@rev.desc.d3.uiv.p1.C.context-d1.share-agree-stable\endcsname{76.0}
\expandafter\def\csname ppt@rev.desc.d3.uiv.p1.C.context-d1.share-agree-stable\endcsname{$+$76.0}
\expandafter\def\csname val@rev.desc.d3.uiv.p1.C.dual-d2.N-agree-stable\endcsname{66}
\expandafter\def\csname cnt@rev.desc.d3.uiv.p1.C.dual-d2.N-agree-stable\endcsname{66}
\expandafter\def\csname pct@rev.desc.d3.uiv.p1.C.dual-d2.N-agree-stable\endcsname{6600.0}
\expandafter\def\csname ppt@rev.desc.d3.uiv.p1.C.dual-d2.N-agree-stable\endcsname{$+$6600.0}
\expandafter\def\csname val@rev.desc.d3.uiv.p1.C.dual-d2.N-disagree\endcsname{20}
\expandafter\def\csname cnt@rev.desc.d3.uiv.p1.C.dual-d2.N-disagree\endcsname{20}
\expandafter\def\csname pct@rev.desc.d3.uiv.p1.C.dual-d2.N-disagree\endcsname{2000.0}
\expandafter\def\csname ppt@rev.desc.d3.uiv.p1.C.dual-d2.N-disagree\endcsname{$+$2000.0}
\expandafter\def\csname val@rev.desc.d3.uiv.p1.C.dual-d2.boot-resid.ci95.0\endcsname{0.000}
\expandafter\def\csname cnt@rev.desc.d3.uiv.p1.C.dual-d2.boot-resid.ci95.0\endcsname{0}
\expandafter\def\csname pct@rev.desc.d3.uiv.p1.C.dual-d2.boot-resid.ci95.0\endcsname{0.0}
\expandafter\def\csname ppt@rev.desc.d3.uiv.p1.C.dual-d2.boot-resid.ci95.0\endcsname{0.0}
\expandafter\def\csname val@rev.desc.d3.uiv.p1.C.dual-d2.boot-resid.ci95.1\endcsname{0.050}
\expandafter\def\csname cnt@rev.desc.d3.uiv.p1.C.dual-d2.boot-resid.ci95.1\endcsname{0}
\expandafter\def\csname pct@rev.desc.d3.uiv.p1.C.dual-d2.boot-resid.ci95.1\endcsname{5.0}
\expandafter\def\csname ppt@rev.desc.d3.uiv.p1.C.dual-d2.boot-resid.ci95.1\endcsname{$+$5.0}
\expandafter\def\csname val@rev.desc.d3.uiv.p1.C.dual-d2.boot-resid.n-apps\endcsname{12}
\expandafter\def\csname cnt@rev.desc.d3.uiv.p1.C.dual-d2.boot-resid.n-apps\endcsname{12}
\expandafter\def\csname pct@rev.desc.d3.uiv.p1.C.dual-d2.boot-resid.n-apps\endcsname{1200.0}
\expandafter\def\csname ppt@rev.desc.d3.uiv.p1.C.dual-d2.boot-resid.n-apps\endcsname{$+$1200.0}
\expandafter\def\csname val@rev.desc.d3.uiv.p1.C.dual-d2.boot-resid.n-boot-valid\endcsname{4000}
\expandafter\def\csname cnt@rev.desc.d3.uiv.p1.C.dual-d2.boot-resid.n-boot-valid\endcsname{4{,}000}
\expandafter\def\csname pct@rev.desc.d3.uiv.p1.C.dual-d2.boot-resid.n-boot-valid\endcsname{400000.0}
\expandafter\def\csname ppt@rev.desc.d3.uiv.p1.C.dual-d2.boot-resid.n-boot-valid\endcsname{$+$400000.0}
\expandafter\def\csname val@rev.desc.d3.uiv.p1.C.dual-d2.boot-resid.value\endcsname{0.015}
\expandafter\def\csname cnt@rev.desc.d3.uiv.p1.C.dual-d2.boot-resid.value\endcsname{0}
\expandafter\def\csname pct@rev.desc.d3.uiv.p1.C.dual-d2.boot-resid.value\endcsname{1.5}
\expandafter\def\csname ppt@rev.desc.d3.uiv.p1.C.dual-d2.boot-resid.value\endcsname{$+$1.5}
\expandafter\def\csname val@rev.desc.d3.uiv.p1.C.dual-d2.boot-share.ci95.0\endcsname{0.625}
\expandafter\def\csname cnt@rev.desc.d3.uiv.p1.C.dual-d2.boot-share.ci95.0\endcsname{1}
\expandafter\def\csname pct@rev.desc.d3.uiv.p1.C.dual-d2.boot-share.ci95.0\endcsname{62.5}
\expandafter\def\csname ppt@rev.desc.d3.uiv.p1.C.dual-d2.boot-share.ci95.0\endcsname{$+$62.5}
\expandafter\def\csname val@rev.desc.d3.uiv.p1.C.dual-d2.boot-share.ci95.1\endcsname{0.750}
\expandafter\def\csname cnt@rev.desc.d3.uiv.p1.C.dual-d2.boot-share.ci95.1\endcsname{1}
\expandafter\def\csname pct@rev.desc.d3.uiv.p1.C.dual-d2.boot-share.ci95.1\endcsname{75.0}
\expandafter\def\csname ppt@rev.desc.d3.uiv.p1.C.dual-d2.boot-share.ci95.1\endcsname{$+$75.0}
\expandafter\def\csname val@rev.desc.d3.uiv.p1.C.dual-d2.boot-share.n-apps\endcsname{12}
\expandafter\def\csname cnt@rev.desc.d3.uiv.p1.C.dual-d2.boot-share.n-apps\endcsname{12}
\expandafter\def\csname pct@rev.desc.d3.uiv.p1.C.dual-d2.boot-share.n-apps\endcsname{1200.0}
\expandafter\def\csname ppt@rev.desc.d3.uiv.p1.C.dual-d2.boot-share.n-apps\endcsname{$+$1200.0}
\expandafter\def\csname val@rev.desc.d3.uiv.p1.C.dual-d2.boot-share.n-boot-valid\endcsname{4000}
\expandafter\def\csname cnt@rev.desc.d3.uiv.p1.C.dual-d2.boot-share.n-boot-valid\endcsname{4{,}000}
\expandafter\def\csname pct@rev.desc.d3.uiv.p1.C.dual-d2.boot-share.n-boot-valid\endcsname{400000.0}
\expandafter\def\csname ppt@rev.desc.d3.uiv.p1.C.dual-d2.boot-share.n-boot-valid\endcsname{$+$400000.0}
\expandafter\def\csname val@rev.desc.d3.uiv.p1.C.dual-d2.boot-share.value\endcsname{0.688}
\expandafter\def\csname cnt@rev.desc.d3.uiv.p1.C.dual-d2.boot-share.value\endcsname{1}
\expandafter\def\csname pct@rev.desc.d3.uiv.p1.C.dual-d2.boot-share.value\endcsname{68.8}
\expandafter\def\csname ppt@rev.desc.d3.uiv.p1.C.dual-d2.boot-share.value\endcsname{$+$68.8}
\expandafter\def\csname val@rev.desc.d3.uiv.p1.C.dual-d2.n\endcsname{96}
\expandafter\def\csname cnt@rev.desc.d3.uiv.p1.C.dual-d2.n\endcsname{96}
\expandafter\def\csname pct@rev.desc.d3.uiv.p1.C.dual-d2.n\endcsname{9600.0}
\expandafter\def\csname ppt@rev.desc.d3.uiv.p1.C.dual-d2.n\endcsname{$+$9600.0}
\expandafter\def\csname val@rev.desc.d3.uiv.p1.C.dual-d2.reproduces-frozen\endcsname{1}
\expandafter\def\csname cnt@rev.desc.d3.uiv.p1.C.dual-d2.reproduces-frozen\endcsname{1}
\expandafter\def\csname pct@rev.desc.d3.uiv.p1.C.dual-d2.reproduces-frozen\endcsname{100.0}
\expandafter\def\csname ppt@rev.desc.d3.uiv.p1.C.dual-d2.reproduces-frozen\endcsname{$+$100.0}
\expandafter\def\csname val@rev.desc.d3.uiv.p1.C.dual-d2.resid-agree-stable\endcsname{0.015}
\expandafter\def\csname cnt@rev.desc.d3.uiv.p1.C.dual-d2.resid-agree-stable\endcsname{0}
\expandafter\def\csname pct@rev.desc.d3.uiv.p1.C.dual-d2.resid-agree-stable\endcsname{1.5}
\expandafter\def\csname ppt@rev.desc.d3.uiv.p1.C.dual-d2.resid-agree-stable\endcsname{$+$1.5}
\expandafter\def\csname val@rev.desc.d3.uiv.p1.C.dual-d2.resid-disagree\endcsname{0.300}
\expandafter\def\csname cnt@rev.desc.d3.uiv.p1.C.dual-d2.resid-disagree\endcsname{0}
\expandafter\def\csname pct@rev.desc.d3.uiv.p1.C.dual-d2.resid-disagree\endcsname{30.0}
\expandafter\def\csname ppt@rev.desc.d3.uiv.p1.C.dual-d2.resid-disagree\endcsname{$+$30.0}
\expandafter\def\csname val@rev.desc.d3.uiv.p1.C.dual-d2.share-agree-stable\endcsname{0.688}
\expandafter\def\csname cnt@rev.desc.d3.uiv.p1.C.dual-d2.share-agree-stable\endcsname{1}
\expandafter\def\csname pct@rev.desc.d3.uiv.p1.C.dual-d2.share-agree-stable\endcsname{68.8}
\expandafter\def\csname ppt@rev.desc.d3.uiv.p1.C.dual-d2.share-agree-stable\endcsname{$+$68.8}
\expandafter\def\csname val@rev.desc.d3.uiv.p1.C.grid2-d3.N-agree-stable\endcsname{62}
\expandafter\def\csname cnt@rev.desc.d3.uiv.p1.C.grid2-d3.N-agree-stable\endcsname{62}
\expandafter\def\csname pct@rev.desc.d3.uiv.p1.C.grid2-d3.N-agree-stable\endcsname{6200.0}
\expandafter\def\csname ppt@rev.desc.d3.uiv.p1.C.grid2-d3.N-agree-stable\endcsname{$+$6200.0}
\expandafter\def\csname val@rev.desc.d3.uiv.p1.C.grid2-d3.N-disagree\endcsname{22}
\expandafter\def\csname cnt@rev.desc.d3.uiv.p1.C.grid2-d3.N-disagree\endcsname{22}
\expandafter\def\csname pct@rev.desc.d3.uiv.p1.C.grid2-d3.N-disagree\endcsname{2200.0}
\expandafter\def\csname ppt@rev.desc.d3.uiv.p1.C.grid2-d3.N-disagree\endcsname{$+$2200.0}
\expandafter\def\csname val@rev.desc.d3.uiv.p1.C.grid2-d3.boot-resid.ci95.0\endcsname{-0.030}
\expandafter\def\csname cnt@rev.desc.d3.uiv.p1.C.grid2-d3.boot-resid.ci95.0\endcsname{0}
\expandafter\def\csname pct@rev.desc.d3.uiv.p1.C.grid2-d3.boot-resid.ci95.0\endcsname{$-$3.0}
\expandafter\def\csname ppt@rev.desc.d3.uiv.p1.C.grid2-d3.boot-resid.ci95.0\endcsname{$-$3.0}
\expandafter\def\csname val@rev.desc.d3.uiv.p1.C.grid2-d3.boot-resid.ci95.1\endcsname{0.093}
\expandafter\def\csname cnt@rev.desc.d3.uiv.p1.C.grid2-d3.boot-resid.ci95.1\endcsname{0}
\expandafter\def\csname pct@rev.desc.d3.uiv.p1.C.grid2-d3.boot-resid.ci95.1\endcsname{9.3}
\expandafter\def\csname ppt@rev.desc.d3.uiv.p1.C.grid2-d3.boot-resid.ci95.1\endcsname{$+$9.3}
\expandafter\def\csname val@rev.desc.d3.uiv.p1.C.grid2-d3.boot-resid.n-apps\endcsname{12}
\expandafter\def\csname cnt@rev.desc.d3.uiv.p1.C.grid2-d3.boot-resid.n-apps\endcsname{12}
\expandafter\def\csname pct@rev.desc.d3.uiv.p1.C.grid2-d3.boot-resid.n-apps\endcsname{1200.0}
\expandafter\def\csname ppt@rev.desc.d3.uiv.p1.C.grid2-d3.boot-resid.n-apps\endcsname{$+$1200.0}
\expandafter\def\csname val@rev.desc.d3.uiv.p1.C.grid2-d3.boot-resid.n-boot-valid\endcsname{4000}
\expandafter\def\csname cnt@rev.desc.d3.uiv.p1.C.grid2-d3.boot-resid.n-boot-valid\endcsname{4{,}000}
\expandafter\def\csname pct@rev.desc.d3.uiv.p1.C.grid2-d3.boot-resid.n-boot-valid\endcsname{400000.0}
\expandafter\def\csname ppt@rev.desc.d3.uiv.p1.C.grid2-d3.boot-resid.n-boot-valid\endcsname{$+$400000.0}
\expandafter\def\csname val@rev.desc.d3.uiv.p1.C.grid2-d3.boot-resid.value\endcsname{0.032}
\expandafter\def\csname cnt@rev.desc.d3.uiv.p1.C.grid2-d3.boot-resid.value\endcsname{0}
\expandafter\def\csname pct@rev.desc.d3.uiv.p1.C.grid2-d3.boot-resid.value\endcsname{3.2}
\expandafter\def\csname ppt@rev.desc.d3.uiv.p1.C.grid2-d3.boot-resid.value\endcsname{$+$3.2}
\expandafter\def\csname val@rev.desc.d3.uiv.p1.C.grid2-d3.boot-share.ci95.0\endcsname{0.542}
\expandafter\def\csname cnt@rev.desc.d3.uiv.p1.C.grid2-d3.boot-share.ci95.0\endcsname{1}
\expandafter\def\csname pct@rev.desc.d3.uiv.p1.C.grid2-d3.boot-share.ci95.0\endcsname{54.2}
\expandafter\def\csname ppt@rev.desc.d3.uiv.p1.C.grid2-d3.boot-share.ci95.0\endcsname{$+$54.2}
\expandafter\def\csname val@rev.desc.d3.uiv.p1.C.grid2-d3.boot-share.ci95.1\endcsname{0.750}
\expandafter\def\csname cnt@rev.desc.d3.uiv.p1.C.grid2-d3.boot-share.ci95.1\endcsname{1}
\expandafter\def\csname pct@rev.desc.d3.uiv.p1.C.grid2-d3.boot-share.ci95.1\endcsname{75.0}
\expandafter\def\csname ppt@rev.desc.d3.uiv.p1.C.grid2-d3.boot-share.ci95.1\endcsname{$+$75.0}
\expandafter\def\csname val@rev.desc.d3.uiv.p1.C.grid2-d3.boot-share.n-apps\endcsname{12}
\expandafter\def\csname cnt@rev.desc.d3.uiv.p1.C.grid2-d3.boot-share.n-apps\endcsname{12}
\expandafter\def\csname pct@rev.desc.d3.uiv.p1.C.grid2-d3.boot-share.n-apps\endcsname{1200.0}
\expandafter\def\csname ppt@rev.desc.d3.uiv.p1.C.grid2-d3.boot-share.n-apps\endcsname{$+$1200.0}
\expandafter\def\csname val@rev.desc.d3.uiv.p1.C.grid2-d3.boot-share.n-boot-valid\endcsname{4000}
\expandafter\def\csname cnt@rev.desc.d3.uiv.p1.C.grid2-d3.boot-share.n-boot-valid\endcsname{4{,}000}
\expandafter\def\csname pct@rev.desc.d3.uiv.p1.C.grid2-d3.boot-share.n-boot-valid\endcsname{400000.0}
\expandafter\def\csname ppt@rev.desc.d3.uiv.p1.C.grid2-d3.boot-share.n-boot-valid\endcsname{$+$400000.0}
\expandafter\def\csname val@rev.desc.d3.uiv.p1.C.grid2-d3.boot-share.value\endcsname{0.646}
\expandafter\def\csname cnt@rev.desc.d3.uiv.p1.C.grid2-d3.boot-share.value\endcsname{1}
\expandafter\def\csname pct@rev.desc.d3.uiv.p1.C.grid2-d3.boot-share.value\endcsname{64.6}
\expandafter\def\csname ppt@rev.desc.d3.uiv.p1.C.grid2-d3.boot-share.value\endcsname{$+$64.6}
\expandafter\def\csname val@rev.desc.d3.uiv.p1.C.grid2-d3.n\endcsname{96}
\expandafter\def\csname cnt@rev.desc.d3.uiv.p1.C.grid2-d3.n\endcsname{96}
\expandafter\def\csname pct@rev.desc.d3.uiv.p1.C.grid2-d3.n\endcsname{9600.0}
\expandafter\def\csname ppt@rev.desc.d3.uiv.p1.C.grid2-d3.n\endcsname{$+$9600.0}
\expandafter\def\csname val@rev.desc.d3.uiv.p1.C.grid2-d3.reproduces-frozen\endcsname{1}
\expandafter\def\csname cnt@rev.desc.d3.uiv.p1.C.grid2-d3.reproduces-frozen\endcsname{1}
\expandafter\def\csname pct@rev.desc.d3.uiv.p1.C.grid2-d3.reproduces-frozen\endcsname{100.0}
\expandafter\def\csname ppt@rev.desc.d3.uiv.p1.C.grid2-d3.reproduces-frozen\endcsname{$+$100.0}
\expandafter\def\csname val@rev.desc.d3.uiv.p1.C.grid2-d3.resid-agree-stable\endcsname{0.032}
\expandafter\def\csname cnt@rev.desc.d3.uiv.p1.C.grid2-d3.resid-agree-stable\endcsname{0}
\expandafter\def\csname pct@rev.desc.d3.uiv.p1.C.grid2-d3.resid-agree-stable\endcsname{3.2}
\expandafter\def\csname ppt@rev.desc.d3.uiv.p1.C.grid2-d3.resid-agree-stable\endcsname{$+$3.2}
\expandafter\def\csname val@rev.desc.d3.uiv.p1.C.grid2-d3.resid-disagree\endcsname{0.364}
\expandafter\def\csname cnt@rev.desc.d3.uiv.p1.C.grid2-d3.resid-disagree\endcsname{0}
\expandafter\def\csname pct@rev.desc.d3.uiv.p1.C.grid2-d3.resid-disagree\endcsname{36.4}
\expandafter\def\csname ppt@rev.desc.d3.uiv.p1.C.grid2-d3.resid-disagree\endcsname{$+$36.4}
\expandafter\def\csname val@rev.desc.d3.uiv.p1.C.grid2-d3.share-agree-stable\endcsname{0.646}
\expandafter\def\csname cnt@rev.desc.d3.uiv.p1.C.grid2-d3.share-agree-stable\endcsname{1}
\expandafter\def\csname pct@rev.desc.d3.uiv.p1.C.grid2-d3.share-agree-stable\endcsname{64.6}
\expandafter\def\csname ppt@rev.desc.d3.uiv.p1.C.grid2-d3.share-agree-stable\endcsname{$+$64.6}
\expandafter\def\csname val@rev.desc.d3.uiv.p1.C.zoomclick-d3.N-agree-stable\endcsname{76}
\expandafter\def\csname cnt@rev.desc.d3.uiv.p1.C.zoomclick-d3.N-agree-stable\endcsname{76}
\expandafter\def\csname pct@rev.desc.d3.uiv.p1.C.zoomclick-d3.N-agree-stable\endcsname{7600.0}
\expandafter\def\csname ppt@rev.desc.d3.uiv.p1.C.zoomclick-d3.N-agree-stable\endcsname{$+$7600.0}
\expandafter\def\csname val@rev.desc.d3.uiv.p1.C.zoomclick-d3.N-disagree\endcsname{13}
\expandafter\def\csname cnt@rev.desc.d3.uiv.p1.C.zoomclick-d3.N-disagree\endcsname{13}
\expandafter\def\csname pct@rev.desc.d3.uiv.p1.C.zoomclick-d3.N-disagree\endcsname{1300.0}
\expandafter\def\csname ppt@rev.desc.d3.uiv.p1.C.zoomclick-d3.N-disagree\endcsname{$+$1300.0}
\expandafter\def\csname val@rev.desc.d3.uiv.p1.C.zoomclick-d3.boot-resid.ci95.0\endcsname{-0.058}
\expandafter\def\csname cnt@rev.desc.d3.uiv.p1.C.zoomclick-d3.boot-resid.ci95.0\endcsname{0}
\expandafter\def\csname pct@rev.desc.d3.uiv.p1.C.zoomclick-d3.boot-resid.ci95.0\endcsname{$-$5.8}
\expandafter\def\csname ppt@rev.desc.d3.uiv.p1.C.zoomclick-d3.boot-resid.ci95.0\endcsname{$-$5.8}
\expandafter\def\csname val@rev.desc.d3.uiv.p1.C.zoomclick-d3.boot-resid.ci95.1\endcsname{0.027}
\expandafter\def\csname cnt@rev.desc.d3.uiv.p1.C.zoomclick-d3.boot-resid.ci95.1\endcsname{0}
\expandafter\def\csname pct@rev.desc.d3.uiv.p1.C.zoomclick-d3.boot-resid.ci95.1\endcsname{2.7}
\expandafter\def\csname ppt@rev.desc.d3.uiv.p1.C.zoomclick-d3.boot-resid.ci95.1\endcsname{$+$2.7}
\expandafter\def\csname val@rev.desc.d3.uiv.p1.C.zoomclick-d3.boot-resid.n-apps\endcsname{12}
\expandafter\def\csname cnt@rev.desc.d3.uiv.p1.C.zoomclick-d3.boot-resid.n-apps\endcsname{12}
\expandafter\def\csname pct@rev.desc.d3.uiv.p1.C.zoomclick-d3.boot-resid.n-apps\endcsname{1200.0}
\expandafter\def\csname ppt@rev.desc.d3.uiv.p1.C.zoomclick-d3.boot-resid.n-apps\endcsname{$+$1200.0}
\expandafter\def\csname val@rev.desc.d3.uiv.p1.C.zoomclick-d3.boot-resid.n-boot-valid\endcsname{4000}
\expandafter\def\csname cnt@rev.desc.d3.uiv.p1.C.zoomclick-d3.boot-resid.n-boot-valid\endcsname{4{,}000}
\expandafter\def\csname pct@rev.desc.d3.uiv.p1.C.zoomclick-d3.boot-resid.n-boot-valid\endcsname{400000.0}
\expandafter\def\csname ppt@rev.desc.d3.uiv.p1.C.zoomclick-d3.boot-resid.n-boot-valid\endcsname{$+$400000.0}
\expandafter\def\csname val@rev.desc.d3.uiv.p1.C.zoomclick-d3.boot-resid.value\endcsname{-0.013}
\expandafter\def\csname cnt@rev.desc.d3.uiv.p1.C.zoomclick-d3.boot-resid.value\endcsname{0}
\expandafter\def\csname pct@rev.desc.d3.uiv.p1.C.zoomclick-d3.boot-resid.value\endcsname{$-$1.3}
\expandafter\def\csname ppt@rev.desc.d3.uiv.p1.C.zoomclick-d3.boot-resid.value\endcsname{$-$1.3}
\expandafter\def\csname val@rev.desc.d3.uiv.p1.C.zoomclick-d3.boot-share.ci95.0\endcsname{0.708}
\expandafter\def\csname cnt@rev.desc.d3.uiv.p1.C.zoomclick-d3.boot-share.ci95.0\endcsname{1}
\expandafter\def\csname pct@rev.desc.d3.uiv.p1.C.zoomclick-d3.boot-share.ci95.0\endcsname{70.8}
\expandafter\def\csname ppt@rev.desc.d3.uiv.p1.C.zoomclick-d3.boot-share.ci95.0\endcsname{$+$70.8}
\expandafter\def\csname val@rev.desc.d3.uiv.p1.C.zoomclick-d3.boot-share.ci95.1\endcsname{0.865}
\expandafter\def\csname cnt@rev.desc.d3.uiv.p1.C.zoomclick-d3.boot-share.ci95.1\endcsname{1}
\expandafter\def\csname pct@rev.desc.d3.uiv.p1.C.zoomclick-d3.boot-share.ci95.1\endcsname{86.5}
\expandafter\def\csname ppt@rev.desc.d3.uiv.p1.C.zoomclick-d3.boot-share.ci95.1\endcsname{$+$86.5}
\expandafter\def\csname val@rev.desc.d3.uiv.p1.C.zoomclick-d3.boot-share.n-apps\endcsname{12}
\expandafter\def\csname cnt@rev.desc.d3.uiv.p1.C.zoomclick-d3.boot-share.n-apps\endcsname{12}
\expandafter\def\csname pct@rev.desc.d3.uiv.p1.C.zoomclick-d3.boot-share.n-apps\endcsname{1200.0}
\expandafter\def\csname ppt@rev.desc.d3.uiv.p1.C.zoomclick-d3.boot-share.n-apps\endcsname{$+$1200.0}
\expandafter\def\csname val@rev.desc.d3.uiv.p1.C.zoomclick-d3.boot-share.n-boot-valid\endcsname{4000}
\expandafter\def\csname cnt@rev.desc.d3.uiv.p1.C.zoomclick-d3.boot-share.n-boot-valid\endcsname{4{,}000}
\expandafter\def\csname pct@rev.desc.d3.uiv.p1.C.zoomclick-d3.boot-share.n-boot-valid\endcsname{400000.0}
\expandafter\def\csname ppt@rev.desc.d3.uiv.p1.C.zoomclick-d3.boot-share.n-boot-valid\endcsname{$+$400000.0}
\expandafter\def\csname val@rev.desc.d3.uiv.p1.C.zoomclick-d3.boot-share.value\endcsname{0.792}
\expandafter\def\csname cnt@rev.desc.d3.uiv.p1.C.zoomclick-d3.boot-share.value\endcsname{1}
\expandafter\def\csname pct@rev.desc.d3.uiv.p1.C.zoomclick-d3.boot-share.value\endcsname{79.2}
\expandafter\def\csname ppt@rev.desc.d3.uiv.p1.C.zoomclick-d3.boot-share.value\endcsname{$+$79.2}
\expandafter\def\csname val@rev.desc.d3.uiv.p1.C.zoomclick-d3.n\endcsname{96}
\expandafter\def\csname cnt@rev.desc.d3.uiv.p1.C.zoomclick-d3.n\endcsname{96}
\expandafter\def\csname pct@rev.desc.d3.uiv.p1.C.zoomclick-d3.n\endcsname{9600.0}
\expandafter\def\csname ppt@rev.desc.d3.uiv.p1.C.zoomclick-d3.n\endcsname{$+$9600.0}
\expandafter\def\csname val@rev.desc.d3.uiv.p1.C.zoomclick-d3.reproduces-frozen\endcsname{1}
\expandafter\def\csname cnt@rev.desc.d3.uiv.p1.C.zoomclick-d3.reproduces-frozen\endcsname{1}
\expandafter\def\csname pct@rev.desc.d3.uiv.p1.C.zoomclick-d3.reproduces-frozen\endcsname{100.0}
\expandafter\def\csname ppt@rev.desc.d3.uiv.p1.C.zoomclick-d3.reproduces-frozen\endcsname{$+$100.0}
\expandafter\def\csname val@rev.desc.d3.uiv.p1.C.zoomclick-d3.resid-agree-stable\endcsname{-0.013}
\expandafter\def\csname cnt@rev.desc.d3.uiv.p1.C.zoomclick-d3.resid-agree-stable\endcsname{0}
\expandafter\def\csname pct@rev.desc.d3.uiv.p1.C.zoomclick-d3.resid-agree-stable\endcsname{$-$1.3}
\expandafter\def\csname ppt@rev.desc.d3.uiv.p1.C.zoomclick-d3.resid-agree-stable\endcsname{$-$1.3}
\expandafter\def\csname val@rev.desc.d3.uiv.p1.C.zoomclick-d3.resid-disagree\endcsname{0.154}
\expandafter\def\csname cnt@rev.desc.d3.uiv.p1.C.zoomclick-d3.resid-disagree\endcsname{0}
\expandafter\def\csname pct@rev.desc.d3.uiv.p1.C.zoomclick-d3.resid-disagree\endcsname{15.4}
\expandafter\def\csname ppt@rev.desc.d3.uiv.p1.C.zoomclick-d3.resid-disagree\endcsname{$+$15.4}
\expandafter\def\csname val@rev.desc.d3.uiv.p1.C.zoomclick-d3.share-agree-stable\endcsname{0.792}
\expandafter\def\csname cnt@rev.desc.d3.uiv.p1.C.zoomclick-d3.share-agree-stable\endcsname{1}
\expandafter\def\csname pct@rev.desc.d3.uiv.p1.C.zoomclick-d3.share-agree-stable\endcsname{79.2}
\expandafter\def\csname ppt@rev.desc.d3.uiv.p1.C.zoomclick-d3.share-agree-stable\endcsname{$+$79.2}
\expandafter\def\csname val@rev.desc.d3.uiv.p2.C.catalog-pooled.N-agree-stable\endcsname{334}
\expandafter\def\csname cnt@rev.desc.d3.uiv.p2.C.catalog-pooled.N-agree-stable\endcsname{334}
\expandafter\def\csname pct@rev.desc.d3.uiv.p2.C.catalog-pooled.N-agree-stable\endcsname{33400.0}
\expandafter\def\csname ppt@rev.desc.d3.uiv.p2.C.catalog-pooled.N-agree-stable\endcsname{$+$33400.0}
\expandafter\def\csname val@rev.desc.d3.uiv.p2.C.catalog-pooled.N-disagree\endcsname{39}
\expandafter\def\csname cnt@rev.desc.d3.uiv.p2.C.catalog-pooled.N-disagree\endcsname{39}
\expandafter\def\csname pct@rev.desc.d3.uiv.p2.C.catalog-pooled.N-disagree\endcsname{3900.0}
\expandafter\def\csname ppt@rev.desc.d3.uiv.p2.C.catalog-pooled.N-disagree\endcsname{$+$3900.0}
\expandafter\def\csname val@rev.desc.d3.uiv.p2.C.catalog-pooled.boot-resid.ci95.0\endcsname{-0.030}
\expandafter\def\csname cnt@rev.desc.d3.uiv.p2.C.catalog-pooled.boot-resid.ci95.0\endcsname{0}
\expandafter\def\csname pct@rev.desc.d3.uiv.p2.C.catalog-pooled.boot-resid.ci95.0\endcsname{$-$3.0}
\expandafter\def\csname ppt@rev.desc.d3.uiv.p2.C.catalog-pooled.boot-resid.ci95.0\endcsname{$-$3.0}
\expandafter\def\csname val@rev.desc.d3.uiv.p2.C.catalog-pooled.boot-resid.ci95.1\endcsname{0.043}
\expandafter\def\csname cnt@rev.desc.d3.uiv.p2.C.catalog-pooled.boot-resid.ci95.1\endcsname{0}
\expandafter\def\csname pct@rev.desc.d3.uiv.p2.C.catalog-pooled.boot-resid.ci95.1\endcsname{4.3}
\expandafter\def\csname ppt@rev.desc.d3.uiv.p2.C.catalog-pooled.boot-resid.ci95.1\endcsname{$+$4.3}
\expandafter\def\csname val@rev.desc.d3.uiv.p2.C.catalog-pooled.boot-resid.n-apps\endcsname{12}
\expandafter\def\csname cnt@rev.desc.d3.uiv.p2.C.catalog-pooled.boot-resid.n-apps\endcsname{12}
\expandafter\def\csname pct@rev.desc.d3.uiv.p2.C.catalog-pooled.boot-resid.n-apps\endcsname{1200.0}
\expandafter\def\csname ppt@rev.desc.d3.uiv.p2.C.catalog-pooled.boot-resid.n-apps\endcsname{$+$1200.0}
\expandafter\def\csname val@rev.desc.d3.uiv.p2.C.catalog-pooled.boot-resid.n-boot-valid\endcsname{4000}
\expandafter\def\csname cnt@rev.desc.d3.uiv.p2.C.catalog-pooled.boot-resid.n-boot-valid\endcsname{4{,}000}
\expandafter\def\csname pct@rev.desc.d3.uiv.p2.C.catalog-pooled.boot-resid.n-boot-valid\endcsname{400000.0}
\expandafter\def\csname ppt@rev.desc.d3.uiv.p2.C.catalog-pooled.boot-resid.n-boot-valid\endcsname{$+$400000.0}
\expandafter\def\csname val@rev.desc.d3.uiv.p2.C.catalog-pooled.boot-resid.value\endcsname{0.006}
\expandafter\def\csname cnt@rev.desc.d3.uiv.p2.C.catalog-pooled.boot-resid.value\endcsname{0}
\expandafter\def\csname pct@rev.desc.d3.uiv.p2.C.catalog-pooled.boot-resid.value\endcsname{0.6}
\expandafter\def\csname ppt@rev.desc.d3.uiv.p2.C.catalog-pooled.boot-resid.value\endcsname{$+$0.6}
\expandafter\def\csname val@rev.desc.d3.uiv.p2.C.catalog-pooled.boot-share.ci95.0\endcsname{0.815}
\expandafter\def\csname cnt@rev.desc.d3.uiv.p2.C.catalog-pooled.boot-share.ci95.0\endcsname{1}
\expandafter\def\csname pct@rev.desc.d3.uiv.p2.C.catalog-pooled.boot-share.ci95.0\endcsname{81.5}
\expandafter\def\csname ppt@rev.desc.d3.uiv.p2.C.catalog-pooled.boot-share.ci95.0\endcsname{$+$81.5}
\expandafter\def\csname val@rev.desc.d3.uiv.p2.C.catalog-pooled.boot-share.ci95.1\endcsname{0.919}
\expandafter\def\csname cnt@rev.desc.d3.uiv.p2.C.catalog-pooled.boot-share.ci95.1\endcsname{1}
\expandafter\def\csname pct@rev.desc.d3.uiv.p2.C.catalog-pooled.boot-share.ci95.1\endcsname{91.9}
\expandafter\def\csname ppt@rev.desc.d3.uiv.p2.C.catalog-pooled.boot-share.ci95.1\endcsname{$+$91.9}
\expandafter\def\csname val@rev.desc.d3.uiv.p2.C.catalog-pooled.boot-share.n-apps\endcsname{12}
\expandafter\def\csname cnt@rev.desc.d3.uiv.p2.C.catalog-pooled.boot-share.n-apps\endcsname{12}
\expandafter\def\csname pct@rev.desc.d3.uiv.p2.C.catalog-pooled.boot-share.n-apps\endcsname{1200.0}
\expandafter\def\csname ppt@rev.desc.d3.uiv.p2.C.catalog-pooled.boot-share.n-apps\endcsname{$+$1200.0}
\expandafter\def\csname val@rev.desc.d3.uiv.p2.C.catalog-pooled.boot-share.n-boot-valid\endcsname{4000}
\expandafter\def\csname cnt@rev.desc.d3.uiv.p2.C.catalog-pooled.boot-share.n-boot-valid\endcsname{4{,}000}
\expandafter\def\csname pct@rev.desc.d3.uiv.p2.C.catalog-pooled.boot-share.n-boot-valid\endcsname{400000.0}
\expandafter\def\csname ppt@rev.desc.d3.uiv.p2.C.catalog-pooled.boot-share.n-boot-valid\endcsname{$+$400000.0}
\expandafter\def\csname val@rev.desc.d3.uiv.p2.C.catalog-pooled.boot-share.value\endcsname{0.870}
\expandafter\def\csname cnt@rev.desc.d3.uiv.p2.C.catalog-pooled.boot-share.value\endcsname{1}
\expandafter\def\csname pct@rev.desc.d3.uiv.p2.C.catalog-pooled.boot-share.value\endcsname{87.0}
\expandafter\def\csname ppt@rev.desc.d3.uiv.p2.C.catalog-pooled.boot-share.value\endcsname{$+$87.0}
\expandafter\def\csname val@rev.desc.d3.uiv.p2.C.catalog-pooled.gap\endcsname{0.020}
\expandafter\def\csname cnt@rev.desc.d3.uiv.p2.C.catalog-pooled.gap\endcsname{0}
\expandafter\def\csname pct@rev.desc.d3.uiv.p2.C.catalog-pooled.gap\endcsname{2.0}
\expandafter\def\csname ppt@rev.desc.d3.uiv.p2.C.catalog-pooled.gap\endcsname{$+$2.0}
\expandafter\def\csname val@rev.desc.d3.uiv.p2.C.catalog-pooled.n\endcsname{384}
\expandafter\def\csname cnt@rev.desc.d3.uiv.p2.C.catalog-pooled.n\endcsname{384}
\expandafter\def\csname pct@rev.desc.d3.uiv.p2.C.catalog-pooled.n\endcsname{38400.0}
\expandafter\def\csname ppt@rev.desc.d3.uiv.p2.C.catalog-pooled.n\endcsname{$+$38400.0}
\expandafter\def\csname val@rev.desc.d3.uiv.p2.C.catalog-pooled.reproduces-frozen\endcsname{1}
\expandafter\def\csname cnt@rev.desc.d3.uiv.p2.C.catalog-pooled.reproduces-frozen\endcsname{1}
\expandafter\def\csname pct@rev.desc.d3.uiv.p2.C.catalog-pooled.reproduces-frozen\endcsname{100.0}
\expandafter\def\csname ppt@rev.desc.d3.uiv.p2.C.catalog-pooled.reproduces-frozen\endcsname{$+$100.0}
\expandafter\def\csname val@rev.desc.d3.uiv.p2.C.catalog-pooled.resid-agree-stable\endcsname{0.006}
\expandafter\def\csname cnt@rev.desc.d3.uiv.p2.C.catalog-pooled.resid-agree-stable\endcsname{0}
\expandafter\def\csname pct@rev.desc.d3.uiv.p2.C.catalog-pooled.resid-agree-stable\endcsname{0.6}
\expandafter\def\csname ppt@rev.desc.d3.uiv.p2.C.catalog-pooled.resid-agree-stable\endcsname{$+$0.6}
\expandafter\def\csname val@rev.desc.d3.uiv.p2.C.catalog-pooled.resid-disagree\endcsname{0.026}
\expandafter\def\csname cnt@rev.desc.d3.uiv.p2.C.catalog-pooled.resid-disagree\endcsname{0}
\expandafter\def\csname pct@rev.desc.d3.uiv.p2.C.catalog-pooled.resid-disagree\endcsname{2.6}
\expandafter\def\csname ppt@rev.desc.d3.uiv.p2.C.catalog-pooled.resid-disagree\endcsname{$+$2.6}
\expandafter\def\csname val@rev.desc.d3.uiv.p2.C.catalog-pooled.share-agree-stable\endcsname{0.870}
\expandafter\def\csname cnt@rev.desc.d3.uiv.p2.C.catalog-pooled.share-agree-stable\endcsname{1}
\expandafter\def\csname pct@rev.desc.d3.uiv.p2.C.catalog-pooled.share-agree-stable\endcsname{87.0}
\expandafter\def\csname ppt@rev.desc.d3.uiv.p2.C.catalog-pooled.share-agree-stable\endcsname{$+$87.0}
\expandafter\def\csname val@rev.desc.d3.uiv.p2.C.center-d1.N-agree-stable\endcsname{87}
\expandafter\def\csname cnt@rev.desc.d3.uiv.p2.C.center-d1.N-agree-stable\endcsname{87}
\expandafter\def\csname pct@rev.desc.d3.uiv.p2.C.center-d1.N-agree-stable\endcsname{8700.0}
\expandafter\def\csname ppt@rev.desc.d3.uiv.p2.C.center-d1.N-agree-stable\endcsname{$+$8700.0}
\expandafter\def\csname val@rev.desc.d3.uiv.p2.C.center-d1.N-disagree\endcsname{8}
\expandafter\def\csname cnt@rev.desc.d3.uiv.p2.C.center-d1.N-disagree\endcsname{8}
\expandafter\def\csname pct@rev.desc.d3.uiv.p2.C.center-d1.N-disagree\endcsname{800.0}
\expandafter\def\csname ppt@rev.desc.d3.uiv.p2.C.center-d1.N-disagree\endcsname{$+$800.0}
\expandafter\def\csname val@rev.desc.d3.uiv.p2.C.center-d1.boot-resid.ci95.0\endcsname{-0.033}
\expandafter\def\csname cnt@rev.desc.d3.uiv.p2.C.center-d1.boot-resid.ci95.0\endcsname{0}
\expandafter\def\csname pct@rev.desc.d3.uiv.p2.C.center-d1.boot-resid.ci95.0\endcsname{$-$3.3}
\expandafter\def\csname ppt@rev.desc.d3.uiv.p2.C.center-d1.boot-resid.ci95.0\endcsname{$-$3.3}
\expandafter\def\csname val@rev.desc.d3.uiv.p2.C.center-d1.boot-resid.ci95.1\endcsname{0.036}
\expandafter\def\csname cnt@rev.desc.d3.uiv.p2.C.center-d1.boot-resid.ci95.1\endcsname{0}
\expandafter\def\csname pct@rev.desc.d3.uiv.p2.C.center-d1.boot-resid.ci95.1\endcsname{3.6}
\expandafter\def\csname ppt@rev.desc.d3.uiv.p2.C.center-d1.boot-resid.ci95.1\endcsname{$+$3.6}
\expandafter\def\csname val@rev.desc.d3.uiv.p2.C.center-d1.boot-resid.n-apps\endcsname{12}
\expandafter\def\csname cnt@rev.desc.d3.uiv.p2.C.center-d1.boot-resid.n-apps\endcsname{12}
\expandafter\def\csname pct@rev.desc.d3.uiv.p2.C.center-d1.boot-resid.n-apps\endcsname{1200.0}
\expandafter\def\csname ppt@rev.desc.d3.uiv.p2.C.center-d1.boot-resid.n-apps\endcsname{$+$1200.0}
\expandafter\def\csname val@rev.desc.d3.uiv.p2.C.center-d1.boot-resid.n-boot-valid\endcsname{4000}
\expandafter\def\csname cnt@rev.desc.d3.uiv.p2.C.center-d1.boot-resid.n-boot-valid\endcsname{4{,}000}
\expandafter\def\csname pct@rev.desc.d3.uiv.p2.C.center-d1.boot-resid.n-boot-valid\endcsname{400000.0}
\expandafter\def\csname ppt@rev.desc.d3.uiv.p2.C.center-d1.boot-resid.n-boot-valid\endcsname{$+$400000.0}
\expandafter\def\csname val@rev.desc.d3.uiv.p2.C.center-d1.boot-resid.value\endcsname{0.000}
\expandafter\def\csname cnt@rev.desc.d3.uiv.p2.C.center-d1.boot-resid.value\endcsname{0}
\expandafter\def\csname pct@rev.desc.d3.uiv.p2.C.center-d1.boot-resid.value\endcsname{0.0}
\expandafter\def\csname ppt@rev.desc.d3.uiv.p2.C.center-d1.boot-resid.value\endcsname{0.0}
\expandafter\def\csname val@rev.desc.d3.uiv.p2.C.center-d1.boot-share.ci95.0\endcsname{0.854}
\expandafter\def\csname cnt@rev.desc.d3.uiv.p2.C.center-d1.boot-share.ci95.0\endcsname{1}
\expandafter\def\csname pct@rev.desc.d3.uiv.p2.C.center-d1.boot-share.ci95.0\endcsname{85.4}
\expandafter\def\csname ppt@rev.desc.d3.uiv.p2.C.center-d1.boot-share.ci95.0\endcsname{$+$85.4}
\expandafter\def\csname val@rev.desc.d3.uiv.p2.C.center-d1.boot-share.ci95.1\endcsname{0.958}
\expandafter\def\csname cnt@rev.desc.d3.uiv.p2.C.center-d1.boot-share.ci95.1\endcsname{1}
\expandafter\def\csname pct@rev.desc.d3.uiv.p2.C.center-d1.boot-share.ci95.1\endcsname{95.8}
\expandafter\def\csname ppt@rev.desc.d3.uiv.p2.C.center-d1.boot-share.ci95.1\endcsname{$+$95.8}
\expandafter\def\csname val@rev.desc.d3.uiv.p2.C.center-d1.boot-share.n-apps\endcsname{12}
\expandafter\def\csname cnt@rev.desc.d3.uiv.p2.C.center-d1.boot-share.n-apps\endcsname{12}
\expandafter\def\csname pct@rev.desc.d3.uiv.p2.C.center-d1.boot-share.n-apps\endcsname{1200.0}
\expandafter\def\csname ppt@rev.desc.d3.uiv.p2.C.center-d1.boot-share.n-apps\endcsname{$+$1200.0}
\expandafter\def\csname val@rev.desc.d3.uiv.p2.C.center-d1.boot-share.n-boot-valid\endcsname{4000}
\expandafter\def\csname cnt@rev.desc.d3.uiv.p2.C.center-d1.boot-share.n-boot-valid\endcsname{4{,}000}
\expandafter\def\csname pct@rev.desc.d3.uiv.p2.C.center-d1.boot-share.n-boot-valid\endcsname{400000.0}
\expandafter\def\csname ppt@rev.desc.d3.uiv.p2.C.center-d1.boot-share.n-boot-valid\endcsname{$+$400000.0}
\expandafter\def\csname val@rev.desc.d3.uiv.p2.C.center-d1.boot-share.value\endcsname{0.906}
\expandafter\def\csname cnt@rev.desc.d3.uiv.p2.C.center-d1.boot-share.value\endcsname{1}
\expandafter\def\csname pct@rev.desc.d3.uiv.p2.C.center-d1.boot-share.value\endcsname{90.6}
\expandafter\def\csname ppt@rev.desc.d3.uiv.p2.C.center-d1.boot-share.value\endcsname{$+$90.6}
\expandafter\def\csname val@rev.desc.d3.uiv.p2.C.center-d1.n\endcsname{96}
\expandafter\def\csname cnt@rev.desc.d3.uiv.p2.C.center-d1.n\endcsname{96}
\expandafter\def\csname pct@rev.desc.d3.uiv.p2.C.center-d1.n\endcsname{9600.0}
\expandafter\def\csname ppt@rev.desc.d3.uiv.p2.C.center-d1.n\endcsname{$+$9600.0}
\expandafter\def\csname val@rev.desc.d3.uiv.p2.C.center-d1.reproduces-frozen\endcsname{1}
\expandafter\def\csname cnt@rev.desc.d3.uiv.p2.C.center-d1.reproduces-frozen\endcsname{1}
\expandafter\def\csname pct@rev.desc.d3.uiv.p2.C.center-d1.reproduces-frozen\endcsname{100.0}
\expandafter\def\csname ppt@rev.desc.d3.uiv.p2.C.center-d1.reproduces-frozen\endcsname{$+$100.0}
\expandafter\def\csname val@rev.desc.d3.uiv.p2.C.center-d1.resid-agree-stable\endcsname{0.000}
\expandafter\def\csname cnt@rev.desc.d3.uiv.p2.C.center-d1.resid-agree-stable\endcsname{0}
\expandafter\def\csname pct@rev.desc.d3.uiv.p2.C.center-d1.resid-agree-stable\endcsname{0.0}
\expandafter\def\csname ppt@rev.desc.d3.uiv.p2.C.center-d1.resid-agree-stable\endcsname{0.0}
\expandafter\def\csname val@rev.desc.d3.uiv.p2.C.center-d1.resid-disagree\endcsname{0.125}
\expandafter\def\csname cnt@rev.desc.d3.uiv.p2.C.center-d1.resid-disagree\endcsname{0}
\expandafter\def\csname pct@rev.desc.d3.uiv.p2.C.center-d1.resid-disagree\endcsname{12.5}
\expandafter\def\csname ppt@rev.desc.d3.uiv.p2.C.center-d1.resid-disagree\endcsname{$+$12.5}
\expandafter\def\csname val@rev.desc.d3.uiv.p2.C.center-d1.share-agree-stable\endcsname{0.906}
\expandafter\def\csname cnt@rev.desc.d3.uiv.p2.C.center-d1.share-agree-stable\endcsname{1}
\expandafter\def\csname pct@rev.desc.d3.uiv.p2.C.center-d1.share-agree-stable\endcsname{90.6}
\expandafter\def\csname ppt@rev.desc.d3.uiv.p2.C.center-d1.share-agree-stable\endcsname{$+$90.6}
\expandafter\def\csname val@rev.desc.d3.uiv.p2.C.context-d1.N-agree-stable\endcsname{84}
\expandafter\def\csname cnt@rev.desc.d3.uiv.p2.C.context-d1.N-agree-stable\endcsname{84}
\expandafter\def\csname pct@rev.desc.d3.uiv.p2.C.context-d1.N-agree-stable\endcsname{8400.0}
\expandafter\def\csname ppt@rev.desc.d3.uiv.p2.C.context-d1.N-agree-stable\endcsname{$+$8400.0}
\expandafter\def\csname val@rev.desc.d3.uiv.p2.C.context-d1.N-disagree\endcsname{10}
\expandafter\def\csname cnt@rev.desc.d3.uiv.p2.C.context-d1.N-disagree\endcsname{10}
\expandafter\def\csname pct@rev.desc.d3.uiv.p2.C.context-d1.N-disagree\endcsname{1000.0}
\expandafter\def\csname ppt@rev.desc.d3.uiv.p2.C.context-d1.N-disagree\endcsname{$+$1000.0}
\expandafter\def\csname val@rev.desc.d3.uiv.p2.C.context-d1.boot-resid.ci95.0\endcsname{-0.034}
\expandafter\def\csname cnt@rev.desc.d3.uiv.p2.C.context-d1.boot-resid.ci95.0\endcsname{0}
\expandafter\def\csname pct@rev.desc.d3.uiv.p2.C.context-d1.boot-resid.ci95.0\endcsname{$-$3.4}
\expandafter\def\csname ppt@rev.desc.d3.uiv.p2.C.context-d1.boot-resid.ci95.0\endcsname{$-$3.4}
\expandafter\def\csname val@rev.desc.d3.uiv.p2.C.context-d1.boot-resid.ci95.1\endcsname{0.037}
\expandafter\def\csname cnt@rev.desc.d3.uiv.p2.C.context-d1.boot-resid.ci95.1\endcsname{0}
\expandafter\def\csname pct@rev.desc.d3.uiv.p2.C.context-d1.boot-resid.ci95.1\endcsname{3.7}
\expandafter\def\csname ppt@rev.desc.d3.uiv.p2.C.context-d1.boot-resid.ci95.1\endcsname{$+$3.7}
\expandafter\def\csname val@rev.desc.d3.uiv.p2.C.context-d1.boot-resid.n-apps\endcsname{12}
\expandafter\def\csname cnt@rev.desc.d3.uiv.p2.C.context-d1.boot-resid.n-apps\endcsname{12}
\expandafter\def\csname pct@rev.desc.d3.uiv.p2.C.context-d1.boot-resid.n-apps\endcsname{1200.0}
\expandafter\def\csname ppt@rev.desc.d3.uiv.p2.C.context-d1.boot-resid.n-apps\endcsname{$+$1200.0}
\expandafter\def\csname val@rev.desc.d3.uiv.p2.C.context-d1.boot-resid.n-boot-valid\endcsname{4000}
\expandafter\def\csname cnt@rev.desc.d3.uiv.p2.C.context-d1.boot-resid.n-boot-valid\endcsname{4{,}000}
\expandafter\def\csname pct@rev.desc.d3.uiv.p2.C.context-d1.boot-resid.n-boot-valid\endcsname{400000.0}
\expandafter\def\csname ppt@rev.desc.d3.uiv.p2.C.context-d1.boot-resid.n-boot-valid\endcsname{$+$400000.0}
\expandafter\def\csname val@rev.desc.d3.uiv.p2.C.context-d1.boot-resid.value\endcsname{0.000}
\expandafter\def\csname cnt@rev.desc.d3.uiv.p2.C.context-d1.boot-resid.value\endcsname{0}
\expandafter\def\csname pct@rev.desc.d3.uiv.p2.C.context-d1.boot-resid.value\endcsname{0.0}
\expandafter\def\csname ppt@rev.desc.d3.uiv.p2.C.context-d1.boot-resid.value\endcsname{0.0}
\expandafter\def\csname val@rev.desc.d3.uiv.p2.C.context-d1.boot-share.ci95.0\endcsname{0.812}
\expandafter\def\csname cnt@rev.desc.d3.uiv.p2.C.context-d1.boot-share.ci95.0\endcsname{1}
\expandafter\def\csname pct@rev.desc.d3.uiv.p2.C.context-d1.boot-share.ci95.0\endcsname{81.2}
\expandafter\def\csname ppt@rev.desc.d3.uiv.p2.C.context-d1.boot-share.ci95.0\endcsname{$+$81.2}
\expandafter\def\csname val@rev.desc.d3.uiv.p2.C.context-d1.boot-share.ci95.1\endcsname{0.938}
\expandafter\def\csname cnt@rev.desc.d3.uiv.p2.C.context-d1.boot-share.ci95.1\endcsname{1}
\expandafter\def\csname pct@rev.desc.d3.uiv.p2.C.context-d1.boot-share.ci95.1\endcsname{93.8}
\expandafter\def\csname ppt@rev.desc.d3.uiv.p2.C.context-d1.boot-share.ci95.1\endcsname{$+$93.8}
\expandafter\def\csname val@rev.desc.d3.uiv.p2.C.context-d1.boot-share.n-apps\endcsname{12}
\expandafter\def\csname cnt@rev.desc.d3.uiv.p2.C.context-d1.boot-share.n-apps\endcsname{12}
\expandafter\def\csname pct@rev.desc.d3.uiv.p2.C.context-d1.boot-share.n-apps\endcsname{1200.0}
\expandafter\def\csname ppt@rev.desc.d3.uiv.p2.C.context-d1.boot-share.n-apps\endcsname{$+$1200.0}
\expandafter\def\csname val@rev.desc.d3.uiv.p2.C.context-d1.boot-share.n-boot-valid\endcsname{4000}
\expandafter\def\csname cnt@rev.desc.d3.uiv.p2.C.context-d1.boot-share.n-boot-valid\endcsname{4{,}000}
\expandafter\def\csname pct@rev.desc.d3.uiv.p2.C.context-d1.boot-share.n-boot-valid\endcsname{400000.0}
\expandafter\def\csname ppt@rev.desc.d3.uiv.p2.C.context-d1.boot-share.n-boot-valid\endcsname{$+$400000.0}
\expandafter\def\csname val@rev.desc.d3.uiv.p2.C.context-d1.boot-share.value\endcsname{0.875}
\expandafter\def\csname cnt@rev.desc.d3.uiv.p2.C.context-d1.boot-share.value\endcsname{1}
\expandafter\def\csname pct@rev.desc.d3.uiv.p2.C.context-d1.boot-share.value\endcsname{87.5}
\expandafter\def\csname ppt@rev.desc.d3.uiv.p2.C.context-d1.boot-share.value\endcsname{$+$87.5}
\expandafter\def\csname val@rev.desc.d3.uiv.p2.C.context-d1.n\endcsname{96}
\expandafter\def\csname cnt@rev.desc.d3.uiv.p2.C.context-d1.n\endcsname{96}
\expandafter\def\csname pct@rev.desc.d3.uiv.p2.C.context-d1.n\endcsname{9600.0}
\expandafter\def\csname ppt@rev.desc.d3.uiv.p2.C.context-d1.n\endcsname{$+$9600.0}
\expandafter\def\csname val@rev.desc.d3.uiv.p2.C.context-d1.reproduces-frozen\endcsname{1}
\expandafter\def\csname cnt@rev.desc.d3.uiv.p2.C.context-d1.reproduces-frozen\endcsname{1}
\expandafter\def\csname pct@rev.desc.d3.uiv.p2.C.context-d1.reproduces-frozen\endcsname{100.0}
\expandafter\def\csname ppt@rev.desc.d3.uiv.p2.C.context-d1.reproduces-frozen\endcsname{$+$100.0}
\expandafter\def\csname val@rev.desc.d3.uiv.p2.C.context-d1.resid-agree-stable\endcsname{0.000}
\expandafter\def\csname cnt@rev.desc.d3.uiv.p2.C.context-d1.resid-agree-stable\endcsname{0}
\expandafter\def\csname pct@rev.desc.d3.uiv.p2.C.context-d1.resid-agree-stable\endcsname{0.0}
\expandafter\def\csname ppt@rev.desc.d3.uiv.p2.C.context-d1.resid-agree-stable\endcsname{0.0}
\expandafter\def\csname val@rev.desc.d3.uiv.p2.C.context-d1.resid-disagree\endcsname{-0.100}
\expandafter\def\csname cnt@rev.desc.d3.uiv.p2.C.context-d1.resid-disagree\endcsname{0}
\expandafter\def\csname pct@rev.desc.d3.uiv.p2.C.context-d1.resid-disagree\endcsname{$-$10.0}
\expandafter\def\csname ppt@rev.desc.d3.uiv.p2.C.context-d1.resid-disagree\endcsname{$-$10.0}
\expandafter\def\csname val@rev.desc.d3.uiv.p2.C.context-d1.share-agree-stable\endcsname{0.875}
\expandafter\def\csname cnt@rev.desc.d3.uiv.p2.C.context-d1.share-agree-stable\endcsname{1}
\expandafter\def\csname pct@rev.desc.d3.uiv.p2.C.context-d1.share-agree-stable\endcsname{87.5}
\expandafter\def\csname ppt@rev.desc.d3.uiv.p2.C.context-d1.share-agree-stable\endcsname{$+$87.5}
\expandafter\def\csname val@rev.desc.d3.uiv.p2.C.dual-d2.N-agree-stable\endcsname{86}
\expandafter\def\csname cnt@rev.desc.d3.uiv.p2.C.dual-d2.N-agree-stable\endcsname{86}
\expandafter\def\csname pct@rev.desc.d3.uiv.p2.C.dual-d2.N-agree-stable\endcsname{8600.0}
\expandafter\def\csname ppt@rev.desc.d3.uiv.p2.C.dual-d2.N-agree-stable\endcsname{$+$8600.0}
\expandafter\def\csname val@rev.desc.d3.uiv.p2.C.dual-d2.N-disagree\endcsname{6}
\expandafter\def\csname cnt@rev.desc.d3.uiv.p2.C.dual-d2.N-disagree\endcsname{6}
\expandafter\def\csname pct@rev.desc.d3.uiv.p2.C.dual-d2.N-disagree\endcsname{600.0}
\expandafter\def\csname ppt@rev.desc.d3.uiv.p2.C.dual-d2.N-disagree\endcsname{$+$600.0}
\expandafter\def\csname val@rev.desc.d3.uiv.p2.C.dual-d2.boot-resid.ci95.0\endcsname{-0.057}
\expandafter\def\csname cnt@rev.desc.d3.uiv.p2.C.dual-d2.boot-resid.ci95.0\endcsname{0}
\expandafter\def\csname pct@rev.desc.d3.uiv.p2.C.dual-d2.boot-resid.ci95.0\endcsname{$-$5.7}
\expandafter\def\csname ppt@rev.desc.d3.uiv.p2.C.dual-d2.boot-resid.ci95.0\endcsname{$-$5.7}
\expandafter\def\csname val@rev.desc.d3.uiv.p2.C.dual-d2.boot-resid.ci95.1\endcsname{0.071}
\expandafter\def\csname cnt@rev.desc.d3.uiv.p2.C.dual-d2.boot-resid.ci95.1\endcsname{0}
\expandafter\def\csname pct@rev.desc.d3.uiv.p2.C.dual-d2.boot-resid.ci95.1\endcsname{7.1}
\expandafter\def\csname ppt@rev.desc.d3.uiv.p2.C.dual-d2.boot-resid.ci95.1\endcsname{$+$7.1}
\expandafter\def\csname val@rev.desc.d3.uiv.p2.C.dual-d2.boot-resid.n-apps\endcsname{12}
\expandafter\def\csname cnt@rev.desc.d3.uiv.p2.C.dual-d2.boot-resid.n-apps\endcsname{12}
\expandafter\def\csname pct@rev.desc.d3.uiv.p2.C.dual-d2.boot-resid.n-apps\endcsname{1200.0}
\expandafter\def\csname ppt@rev.desc.d3.uiv.p2.C.dual-d2.boot-resid.n-apps\endcsname{$+$1200.0}
\expandafter\def\csname val@rev.desc.d3.uiv.p2.C.dual-d2.boot-resid.n-boot-valid\endcsname{4000}
\expandafter\def\csname cnt@rev.desc.d3.uiv.p2.C.dual-d2.boot-resid.n-boot-valid\endcsname{4{,}000}
\expandafter\def\csname pct@rev.desc.d3.uiv.p2.C.dual-d2.boot-resid.n-boot-valid\endcsname{400000.0}
\expandafter\def\csname ppt@rev.desc.d3.uiv.p2.C.dual-d2.boot-resid.n-boot-valid\endcsname{$+$400000.0}
\expandafter\def\csname val@rev.desc.d3.uiv.p2.C.dual-d2.boot-resid.value\endcsname{0.000}
\expandafter\def\csname cnt@rev.desc.d3.uiv.p2.C.dual-d2.boot-resid.value\endcsname{0}
\expandafter\def\csname pct@rev.desc.d3.uiv.p2.C.dual-d2.boot-resid.value\endcsname{0.0}
\expandafter\def\csname ppt@rev.desc.d3.uiv.p2.C.dual-d2.boot-resid.value\endcsname{0.0}
\expandafter\def\csname val@rev.desc.d3.uiv.p2.C.dual-d2.boot-share.ci95.0\endcsname{0.823}
\expandafter\def\csname cnt@rev.desc.d3.uiv.p2.C.dual-d2.boot-share.ci95.0\endcsname{1}
\expandafter\def\csname pct@rev.desc.d3.uiv.p2.C.dual-d2.boot-share.ci95.0\endcsname{82.3}
\expandafter\def\csname ppt@rev.desc.d3.uiv.p2.C.dual-d2.boot-share.ci95.0\endcsname{$+$82.3}
\expandafter\def\csname val@rev.desc.d3.uiv.p2.C.dual-d2.boot-share.ci95.1\endcsname{0.958}
\expandafter\def\csname cnt@rev.desc.d3.uiv.p2.C.dual-d2.boot-share.ci95.1\endcsname{1}
\expandafter\def\csname pct@rev.desc.d3.uiv.p2.C.dual-d2.boot-share.ci95.1\endcsname{95.8}
\expandafter\def\csname ppt@rev.desc.d3.uiv.p2.C.dual-d2.boot-share.ci95.1\endcsname{$+$95.8}
\expandafter\def\csname val@rev.desc.d3.uiv.p2.C.dual-d2.boot-share.n-apps\endcsname{12}
\expandafter\def\csname cnt@rev.desc.d3.uiv.p2.C.dual-d2.boot-share.n-apps\endcsname{12}
\expandafter\def\csname pct@rev.desc.d3.uiv.p2.C.dual-d2.boot-share.n-apps\endcsname{1200.0}
\expandafter\def\csname ppt@rev.desc.d3.uiv.p2.C.dual-d2.boot-share.n-apps\endcsname{$+$1200.0}
\expandafter\def\csname val@rev.desc.d3.uiv.p2.C.dual-d2.boot-share.n-boot-valid\endcsname{4000}
\expandafter\def\csname cnt@rev.desc.d3.uiv.p2.C.dual-d2.boot-share.n-boot-valid\endcsname{4{,}000}
\expandafter\def\csname pct@rev.desc.d3.uiv.p2.C.dual-d2.boot-share.n-boot-valid\endcsname{400000.0}
\expandafter\def\csname ppt@rev.desc.d3.uiv.p2.C.dual-d2.boot-share.n-boot-valid\endcsname{$+$400000.0}
\expandafter\def\csname val@rev.desc.d3.uiv.p2.C.dual-d2.boot-share.value\endcsname{0.896}
\expandafter\def\csname cnt@rev.desc.d3.uiv.p2.C.dual-d2.boot-share.value\endcsname{1}
\expandafter\def\csname pct@rev.desc.d3.uiv.p2.C.dual-d2.boot-share.value\endcsname{89.6}
\expandafter\def\csname ppt@rev.desc.d3.uiv.p2.C.dual-d2.boot-share.value\endcsname{$+$89.6}
\expandafter\def\csname val@rev.desc.d3.uiv.p2.C.dual-d2.n\endcsname{96}
\expandafter\def\csname cnt@rev.desc.d3.uiv.p2.C.dual-d2.n\endcsname{96}
\expandafter\def\csname pct@rev.desc.d3.uiv.p2.C.dual-d2.n\endcsname{9600.0}
\expandafter\def\csname ppt@rev.desc.d3.uiv.p2.C.dual-d2.n\endcsname{$+$9600.0}
\expandafter\def\csname val@rev.desc.d3.uiv.p2.C.dual-d2.reproduces-frozen\endcsname{1}
\expandafter\def\csname cnt@rev.desc.d3.uiv.p2.C.dual-d2.reproduces-frozen\endcsname{1}
\expandafter\def\csname pct@rev.desc.d3.uiv.p2.C.dual-d2.reproduces-frozen\endcsname{100.0}
\expandafter\def\csname ppt@rev.desc.d3.uiv.p2.C.dual-d2.reproduces-frozen\endcsname{$+$100.0}
\expandafter\def\csname val@rev.desc.d3.uiv.p2.C.dual-d2.resid-agree-stable\endcsname{0.000}
\expandafter\def\csname cnt@rev.desc.d3.uiv.p2.C.dual-d2.resid-agree-stable\endcsname{0}
\expandafter\def\csname pct@rev.desc.d3.uiv.p2.C.dual-d2.resid-agree-stable\endcsname{0.0}
\expandafter\def\csname ppt@rev.desc.d3.uiv.p2.C.dual-d2.resid-agree-stable\endcsname{0.0}
\expandafter\def\csname val@rev.desc.d3.uiv.p2.C.dual-d2.resid-disagree\endcsname{-0.167}
\expandafter\def\csname cnt@rev.desc.d3.uiv.p2.C.dual-d2.resid-disagree\endcsname{0}
\expandafter\def\csname pct@rev.desc.d3.uiv.p2.C.dual-d2.resid-disagree\endcsname{$-$16.7}
\expandafter\def\csname ppt@rev.desc.d3.uiv.p2.C.dual-d2.resid-disagree\endcsname{$-$16.7}
\expandafter\def\csname val@rev.desc.d3.uiv.p2.C.dual-d2.share-agree-stable\endcsname{0.896}
\expandafter\def\csname cnt@rev.desc.d3.uiv.p2.C.dual-d2.share-agree-stable\endcsname{1}
\expandafter\def\csname pct@rev.desc.d3.uiv.p2.C.dual-d2.share-agree-stable\endcsname{89.6}
\expandafter\def\csname ppt@rev.desc.d3.uiv.p2.C.dual-d2.share-agree-stable\endcsname{$+$89.6}
\expandafter\def\csname val@rev.desc.d3.uiv.p2.C.grid2-d3.N-agree-stable\endcsname{77}
\expandafter\def\csname cnt@rev.desc.d3.uiv.p2.C.grid2-d3.N-agree-stable\endcsname{77}
\expandafter\def\csname pct@rev.desc.d3.uiv.p2.C.grid2-d3.N-agree-stable\endcsname{7700.0}
\expandafter\def\csname ppt@rev.desc.d3.uiv.p2.C.grid2-d3.N-agree-stable\endcsname{$+$7700.0}
\expandafter\def\csname val@rev.desc.d3.uiv.p2.C.grid2-d3.N-disagree\endcsname{15}
\expandafter\def\csname cnt@rev.desc.d3.uiv.p2.C.grid2-d3.N-disagree\endcsname{15}
\expandafter\def\csname pct@rev.desc.d3.uiv.p2.C.grid2-d3.N-disagree\endcsname{1500.0}
\expandafter\def\csname ppt@rev.desc.d3.uiv.p2.C.grid2-d3.N-disagree\endcsname{$+$1500.0}
\expandafter\def\csname val@rev.desc.d3.uiv.p2.C.grid2-d3.boot-resid.ci95.0\endcsname{-0.026}
\expandafter\def\csname cnt@rev.desc.d3.uiv.p2.C.grid2-d3.boot-resid.ci95.0\endcsname{0}
\expandafter\def\csname pct@rev.desc.d3.uiv.p2.C.grid2-d3.boot-resid.ci95.0\endcsname{$-$2.6}
\expandafter\def\csname ppt@rev.desc.d3.uiv.p2.C.grid2-d3.boot-resid.ci95.0\endcsname{$-$2.6}
\expandafter\def\csname val@rev.desc.d3.uiv.p2.C.grid2-d3.boot-resid.ci95.1\endcsname{0.077}
\expandafter\def\csname cnt@rev.desc.d3.uiv.p2.C.grid2-d3.boot-resid.ci95.1\endcsname{0}
\expandafter\def\csname pct@rev.desc.d3.uiv.p2.C.grid2-d3.boot-resid.ci95.1\endcsname{7.7}
\expandafter\def\csname ppt@rev.desc.d3.uiv.p2.C.grid2-d3.boot-resid.ci95.1\endcsname{$+$7.7}
\expandafter\def\csname val@rev.desc.d3.uiv.p2.C.grid2-d3.boot-resid.n-apps\endcsname{12}
\expandafter\def\csname cnt@rev.desc.d3.uiv.p2.C.grid2-d3.boot-resid.n-apps\endcsname{12}
\expandafter\def\csname pct@rev.desc.d3.uiv.p2.C.grid2-d3.boot-resid.n-apps\endcsname{1200.0}
\expandafter\def\csname ppt@rev.desc.d3.uiv.p2.C.grid2-d3.boot-resid.n-apps\endcsname{$+$1200.0}
\expandafter\def\csname val@rev.desc.d3.uiv.p2.C.grid2-d3.boot-resid.n-boot-valid\endcsname{4000}
\expandafter\def\csname cnt@rev.desc.d3.uiv.p2.C.grid2-d3.boot-resid.n-boot-valid\endcsname{4{,}000}
\expandafter\def\csname pct@rev.desc.d3.uiv.p2.C.grid2-d3.boot-resid.n-boot-valid\endcsname{400000.0}
\expandafter\def\csname ppt@rev.desc.d3.uiv.p2.C.grid2-d3.boot-resid.n-boot-valid\endcsname{$+$400000.0}
\expandafter\def\csname val@rev.desc.d3.uiv.p2.C.grid2-d3.boot-resid.value\endcsname{0.026}
\expandafter\def\csname cnt@rev.desc.d3.uiv.p2.C.grid2-d3.boot-resid.value\endcsname{0}
\expandafter\def\csname pct@rev.desc.d3.uiv.p2.C.grid2-d3.boot-resid.value\endcsname{2.6}
\expandafter\def\csname ppt@rev.desc.d3.uiv.p2.C.grid2-d3.boot-resid.value\endcsname{$+$2.6}
\expandafter\def\csname val@rev.desc.d3.uiv.p2.C.grid2-d3.boot-share.ci95.0\endcsname{0.719}
\expandafter\def\csname cnt@rev.desc.d3.uiv.p2.C.grid2-d3.boot-share.ci95.0\endcsname{1}
\expandafter\def\csname pct@rev.desc.d3.uiv.p2.C.grid2-d3.boot-share.ci95.0\endcsname{71.9}
\expandafter\def\csname ppt@rev.desc.d3.uiv.p2.C.grid2-d3.boot-share.ci95.0\endcsname{$+$71.9}
\expandafter\def\csname val@rev.desc.d3.uiv.p2.C.grid2-d3.boot-share.ci95.1\endcsname{0.875}
\expandafter\def\csname cnt@rev.desc.d3.uiv.p2.C.grid2-d3.boot-share.ci95.1\endcsname{1}
\expandafter\def\csname pct@rev.desc.d3.uiv.p2.C.grid2-d3.boot-share.ci95.1\endcsname{87.5}
\expandafter\def\csname ppt@rev.desc.d3.uiv.p2.C.grid2-d3.boot-share.ci95.1\endcsname{$+$87.5}
\expandafter\def\csname val@rev.desc.d3.uiv.p2.C.grid2-d3.boot-share.n-apps\endcsname{12}
\expandafter\def\csname cnt@rev.desc.d3.uiv.p2.C.grid2-d3.boot-share.n-apps\endcsname{12}
\expandafter\def\csname pct@rev.desc.d3.uiv.p2.C.grid2-d3.boot-share.n-apps\endcsname{1200.0}
\expandafter\def\csname ppt@rev.desc.d3.uiv.p2.C.grid2-d3.boot-share.n-apps\endcsname{$+$1200.0}
\expandafter\def\csname val@rev.desc.d3.uiv.p2.C.grid2-d3.boot-share.n-boot-valid\endcsname{4000}
\expandafter\def\csname cnt@rev.desc.d3.uiv.p2.C.grid2-d3.boot-share.n-boot-valid\endcsname{4{,}000}
\expandafter\def\csname pct@rev.desc.d3.uiv.p2.C.grid2-d3.boot-share.n-boot-valid\endcsname{400000.0}
\expandafter\def\csname ppt@rev.desc.d3.uiv.p2.C.grid2-d3.boot-share.n-boot-valid\endcsname{$+$400000.0}
\expandafter\def\csname val@rev.desc.d3.uiv.p2.C.grid2-d3.boot-share.value\endcsname{0.802}
\expandafter\def\csname cnt@rev.desc.d3.uiv.p2.C.grid2-d3.boot-share.value\endcsname{1}
\expandafter\def\csname pct@rev.desc.d3.uiv.p2.C.grid2-d3.boot-share.value\endcsname{80.2}
\expandafter\def\csname ppt@rev.desc.d3.uiv.p2.C.grid2-d3.boot-share.value\endcsname{$+$80.2}
\expandafter\def\csname val@rev.desc.d3.uiv.p2.C.grid2-d3.n\endcsname{96}
\expandafter\def\csname cnt@rev.desc.d3.uiv.p2.C.grid2-d3.n\endcsname{96}
\expandafter\def\csname pct@rev.desc.d3.uiv.p2.C.grid2-d3.n\endcsname{9600.0}
\expandafter\def\csname ppt@rev.desc.d3.uiv.p2.C.grid2-d3.n\endcsname{$+$9600.0}
\expandafter\def\csname val@rev.desc.d3.uiv.p2.C.grid2-d3.reproduces-frozen\endcsname{1}
\expandafter\def\csname cnt@rev.desc.d3.uiv.p2.C.grid2-d3.reproduces-frozen\endcsname{1}
\expandafter\def\csname pct@rev.desc.d3.uiv.p2.C.grid2-d3.reproduces-frozen\endcsname{100.0}
\expandafter\def\csname ppt@rev.desc.d3.uiv.p2.C.grid2-d3.reproduces-frozen\endcsname{$+$100.0}
\expandafter\def\csname val@rev.desc.d3.uiv.p2.C.grid2-d3.resid-agree-stable\endcsname{0.026}
\expandafter\def\csname cnt@rev.desc.d3.uiv.p2.C.grid2-d3.resid-agree-stable\endcsname{0}
\expandafter\def\csname pct@rev.desc.d3.uiv.p2.C.grid2-d3.resid-agree-stable\endcsname{2.6}
\expandafter\def\csname ppt@rev.desc.d3.uiv.p2.C.grid2-d3.resid-agree-stable\endcsname{$+$2.6}
\expandafter\def\csname val@rev.desc.d3.uiv.p2.C.grid2-d3.resid-disagree\endcsname{0.133}
\expandafter\def\csname cnt@rev.desc.d3.uiv.p2.C.grid2-d3.resid-disagree\endcsname{0}
\expandafter\def\csname pct@rev.desc.d3.uiv.p2.C.grid2-d3.resid-disagree\endcsname{13.3}
\expandafter\def\csname ppt@rev.desc.d3.uiv.p2.C.grid2-d3.resid-disagree\endcsname{$+$13.3}
\expandafter\def\csname val@rev.desc.d3.uiv.p2.C.grid2-d3.share-agree-stable\endcsname{0.802}
\expandafter\def\csname cnt@rev.desc.d3.uiv.p2.C.grid2-d3.share-agree-stable\endcsname{1}
\expandafter\def\csname pct@rev.desc.d3.uiv.p2.C.grid2-d3.share-agree-stable\endcsname{80.2}
\expandafter\def\csname ppt@rev.desc.d3.uiv.p2.C.grid2-d3.share-agree-stable\endcsname{$+$80.2}
\expandafter\def\csname val@rev.desc.d3.uiv.p2.C.zoomclick-d3.N-agree-stable\endcsname{85}
\expandafter\def\csname cnt@rev.desc.d3.uiv.p2.C.zoomclick-d3.N-agree-stable\endcsname{85}
\expandafter\def\csname pct@rev.desc.d3.uiv.p2.C.zoomclick-d3.N-agree-stable\endcsname{8500.0}
\expandafter\def\csname ppt@rev.desc.d3.uiv.p2.C.zoomclick-d3.N-agree-stable\endcsname{$+$8500.0}
\expandafter\def\csname val@rev.desc.d3.uiv.p2.C.zoomclick-d3.N-disagree\endcsname{11}
\expandafter\def\csname cnt@rev.desc.d3.uiv.p2.C.zoomclick-d3.N-disagree\endcsname{11}
\expandafter\def\csname pct@rev.desc.d3.uiv.p2.C.zoomclick-d3.N-disagree\endcsname{1100.0}
\expandafter\def\csname ppt@rev.desc.d3.uiv.p2.C.zoomclick-d3.N-disagree\endcsname{$+$1100.0}
\expandafter\def\csname val@rev.desc.d3.uiv.p2.C.zoomclick-d3.boot-resid.ci95.0\endcsname{-0.022}
\expandafter\def\csname cnt@rev.desc.d3.uiv.p2.C.zoomclick-d3.boot-resid.ci95.0\endcsname{0}
\expandafter\def\csname pct@rev.desc.d3.uiv.p2.C.zoomclick-d3.boot-resid.ci95.0\endcsname{$-$2.2}
\expandafter\def\csname ppt@rev.desc.d3.uiv.p2.C.zoomclick-d3.boot-resid.ci95.0\endcsname{$-$2.2}
\expandafter\def\csname val@rev.desc.d3.uiv.p2.C.zoomclick-d3.boot-resid.ci95.1\endcsname{0.073}
\expandafter\def\csname cnt@rev.desc.d3.uiv.p2.C.zoomclick-d3.boot-resid.ci95.1\endcsname{0}
\expandafter\def\csname pct@rev.desc.d3.uiv.p2.C.zoomclick-d3.boot-resid.ci95.1\endcsname{7.3}
\expandafter\def\csname ppt@rev.desc.d3.uiv.p2.C.zoomclick-d3.boot-resid.ci95.1\endcsname{$+$7.3}
\expandafter\def\csname val@rev.desc.d3.uiv.p2.C.zoomclick-d3.boot-resid.n-apps\endcsname{12}
\expandafter\def\csname cnt@rev.desc.d3.uiv.p2.C.zoomclick-d3.boot-resid.n-apps\endcsname{12}
\expandafter\def\csname pct@rev.desc.d3.uiv.p2.C.zoomclick-d3.boot-resid.n-apps\endcsname{1200.0}
\expandafter\def\csname ppt@rev.desc.d3.uiv.p2.C.zoomclick-d3.boot-resid.n-apps\endcsname{$+$1200.0}
\expandafter\def\csname val@rev.desc.d3.uiv.p2.C.zoomclick-d3.boot-resid.n-boot-valid\endcsname{4000}
\expandafter\def\csname cnt@rev.desc.d3.uiv.p2.C.zoomclick-d3.boot-resid.n-boot-valid\endcsname{4{,}000}
\expandafter\def\csname pct@rev.desc.d3.uiv.p2.C.zoomclick-d3.boot-resid.n-boot-valid\endcsname{400000.0}
\expandafter\def\csname ppt@rev.desc.d3.uiv.p2.C.zoomclick-d3.boot-resid.n-boot-valid\endcsname{$+$400000.0}
\expandafter\def\csname val@rev.desc.d3.uiv.p2.C.zoomclick-d3.boot-resid.value\endcsname{0.024}
\expandafter\def\csname cnt@rev.desc.d3.uiv.p2.C.zoomclick-d3.boot-resid.value\endcsname{0}
\expandafter\def\csname pct@rev.desc.d3.uiv.p2.C.zoomclick-d3.boot-resid.value\endcsname{2.4}
\expandafter\def\csname ppt@rev.desc.d3.uiv.p2.C.zoomclick-d3.boot-resid.value\endcsname{$+$2.4}
\expandafter\def\csname val@rev.desc.d3.uiv.p2.C.zoomclick-d3.boot-share.ci95.0\endcsname{0.823}
\expandafter\def\csname cnt@rev.desc.d3.uiv.p2.C.zoomclick-d3.boot-share.ci95.0\endcsname{1}
\expandafter\def\csname pct@rev.desc.d3.uiv.p2.C.zoomclick-d3.boot-share.ci95.0\endcsname{82.3}
\expandafter\def\csname ppt@rev.desc.d3.uiv.p2.C.zoomclick-d3.boot-share.ci95.0\endcsname{$+$82.3}
\expandafter\def\csname val@rev.desc.d3.uiv.p2.C.zoomclick-d3.boot-share.ci95.1\endcsname{0.948}
\expandafter\def\csname cnt@rev.desc.d3.uiv.p2.C.zoomclick-d3.boot-share.ci95.1\endcsname{1}
\expandafter\def\csname pct@rev.desc.d3.uiv.p2.C.zoomclick-d3.boot-share.ci95.1\endcsname{94.8}
\expandafter\def\csname ppt@rev.desc.d3.uiv.p2.C.zoomclick-d3.boot-share.ci95.1\endcsname{$+$94.8}
\expandafter\def\csname val@rev.desc.d3.uiv.p2.C.zoomclick-d3.boot-share.n-apps\endcsname{12}
\expandafter\def\csname cnt@rev.desc.d3.uiv.p2.C.zoomclick-d3.boot-share.n-apps\endcsname{12}
\expandafter\def\csname pct@rev.desc.d3.uiv.p2.C.zoomclick-d3.boot-share.n-apps\endcsname{1200.0}
\expandafter\def\csname ppt@rev.desc.d3.uiv.p2.C.zoomclick-d3.boot-share.n-apps\endcsname{$+$1200.0}
\expandafter\def\csname val@rev.desc.d3.uiv.p2.C.zoomclick-d3.boot-share.n-boot-valid\endcsname{4000}
\expandafter\def\csname cnt@rev.desc.d3.uiv.p2.C.zoomclick-d3.boot-share.n-boot-valid\endcsname{4{,}000}
\expandafter\def\csname pct@rev.desc.d3.uiv.p2.C.zoomclick-d3.boot-share.n-boot-valid\endcsname{400000.0}
\expandafter\def\csname ppt@rev.desc.d3.uiv.p2.C.zoomclick-d3.boot-share.n-boot-valid\endcsname{$+$400000.0}
\expandafter\def\csname val@rev.desc.d3.uiv.p2.C.zoomclick-d3.boot-share.value\endcsname{0.885}
\expandafter\def\csname cnt@rev.desc.d3.uiv.p2.C.zoomclick-d3.boot-share.value\endcsname{1}
\expandafter\def\csname pct@rev.desc.d3.uiv.p2.C.zoomclick-d3.boot-share.value\endcsname{88.5}
\expandafter\def\csname ppt@rev.desc.d3.uiv.p2.C.zoomclick-d3.boot-share.value\endcsname{$+$88.5}
\expandafter\def\csname val@rev.desc.d3.uiv.p2.C.zoomclick-d3.n\endcsname{96}
\expandafter\def\csname cnt@rev.desc.d3.uiv.p2.C.zoomclick-d3.n\endcsname{96}
\expandafter\def\csname pct@rev.desc.d3.uiv.p2.C.zoomclick-d3.n\endcsname{9600.0}
\expandafter\def\csname ppt@rev.desc.d3.uiv.p2.C.zoomclick-d3.n\endcsname{$+$9600.0}
\expandafter\def\csname val@rev.desc.d3.uiv.p2.C.zoomclick-d3.reproduces-frozen\endcsname{1}
\expandafter\def\csname cnt@rev.desc.d3.uiv.p2.C.zoomclick-d3.reproduces-frozen\endcsname{1}
\expandafter\def\csname pct@rev.desc.d3.uiv.p2.C.zoomclick-d3.reproduces-frozen\endcsname{100.0}
\expandafter\def\csname ppt@rev.desc.d3.uiv.p2.C.zoomclick-d3.reproduces-frozen\endcsname{$+$100.0}
\expandafter\def\csname val@rev.desc.d3.uiv.p2.C.zoomclick-d3.resid-agree-stable\endcsname{0.024}
\expandafter\def\csname cnt@rev.desc.d3.uiv.p2.C.zoomclick-d3.resid-agree-stable\endcsname{0}
\expandafter\def\csname pct@rev.desc.d3.uiv.p2.C.zoomclick-d3.resid-agree-stable\endcsname{2.4}
\expandafter\def\csname ppt@rev.desc.d3.uiv.p2.C.zoomclick-d3.resid-agree-stable\endcsname{$+$2.4}
\expandafter\def\csname val@rev.desc.d3.uiv.p2.C.zoomclick-d3.resid-disagree\endcsname{0.182}
\expandafter\def\csname cnt@rev.desc.d3.uiv.p2.C.zoomclick-d3.resid-disagree\endcsname{0}
\expandafter\def\csname pct@rev.desc.d3.uiv.p2.C.zoomclick-d3.resid-disagree\endcsname{18.2}
\expandafter\def\csname ppt@rev.desc.d3.uiv.p2.C.zoomclick-d3.resid-disagree\endcsname{$+$18.2}
\expandafter\def\csname val@rev.desc.d3.uiv.p2.C.zoomclick-d3.share-agree-stable\endcsname{0.885}
\expandafter\def\csname cnt@rev.desc.d3.uiv.p2.C.zoomclick-d3.share-agree-stable\endcsname{1}
\expandafter\def\csname pct@rev.desc.d3.uiv.p2.C.zoomclick-d3.share-agree-stable\endcsname{88.5}
\expandafter\def\csname ppt@rev.desc.d3.uiv.p2.C.zoomclick-d3.share-agree-stable\endcsname{$+$88.5}
\expandafter\def\csname val@rev.desc.d3.uiv.p2.T.catalog-pooled.N-agree-stable\endcsname{512}
\expandafter\def\csname cnt@rev.desc.d3.uiv.p2.T.catalog-pooled.N-agree-stable\endcsname{512}
\expandafter\def\csname pct@rev.desc.d3.uiv.p2.T.catalog-pooled.N-agree-stable\endcsname{51200.0}
\expandafter\def\csname ppt@rev.desc.d3.uiv.p2.T.catalog-pooled.N-agree-stable\endcsname{$+$51200.0}
\expandafter\def\csname val@rev.desc.d3.uiv.p2.T.catalog-pooled.N-disagree\endcsname{47}
\expandafter\def\csname cnt@rev.desc.d3.uiv.p2.T.catalog-pooled.N-disagree\endcsname{47}
\expandafter\def\csname pct@rev.desc.d3.uiv.p2.T.catalog-pooled.N-disagree\endcsname{4700.0}
\expandafter\def\csname ppt@rev.desc.d3.uiv.p2.T.catalog-pooled.N-disagree\endcsname{$+$4700.0}
\expandafter\def\csname val@rev.desc.d3.uiv.p2.T.catalog-pooled.boot-resid.ci95.0\endcsname{-0.026}
\expandafter\def\csname cnt@rev.desc.d3.uiv.p2.T.catalog-pooled.boot-resid.ci95.0\endcsname{0}
\expandafter\def\csname pct@rev.desc.d3.uiv.p2.T.catalog-pooled.boot-resid.ci95.0\endcsname{$-$2.6}
\expandafter\def\csname ppt@rev.desc.d3.uiv.p2.T.catalog-pooled.boot-resid.ci95.0\endcsname{$-$2.6}
\expandafter\def\csname val@rev.desc.d3.uiv.p2.T.catalog-pooled.boot-resid.ci95.1\endcsname{0.008}
\expandafter\def\csname cnt@rev.desc.d3.uiv.p2.T.catalog-pooled.boot-resid.ci95.1\endcsname{0}
\expandafter\def\csname pct@rev.desc.d3.uiv.p2.T.catalog-pooled.boot-resid.ci95.1\endcsname{0.8}
\expandafter\def\csname ppt@rev.desc.d3.uiv.p2.T.catalog-pooled.boot-resid.ci95.1\endcsname{$+$0.8}
\expandafter\def\csname val@rev.desc.d3.uiv.p2.T.catalog-pooled.boot-resid.n-apps\endcsname{24}
\expandafter\def\csname cnt@rev.desc.d3.uiv.p2.T.catalog-pooled.boot-resid.n-apps\endcsname{24}
\expandafter\def\csname pct@rev.desc.d3.uiv.p2.T.catalog-pooled.boot-resid.n-apps\endcsname{2400.0}
\expandafter\def\csname ppt@rev.desc.d3.uiv.p2.T.catalog-pooled.boot-resid.n-apps\endcsname{$+$2400.0}
\expandafter\def\csname val@rev.desc.d3.uiv.p2.T.catalog-pooled.boot-resid.n-boot-valid\endcsname{4000}
\expandafter\def\csname cnt@rev.desc.d3.uiv.p2.T.catalog-pooled.boot-resid.n-boot-valid\endcsname{4{,}000}
\expandafter\def\csname pct@rev.desc.d3.uiv.p2.T.catalog-pooled.boot-resid.n-boot-valid\endcsname{400000.0}
\expandafter\def\csname ppt@rev.desc.d3.uiv.p2.T.catalog-pooled.boot-resid.n-boot-valid\endcsname{$+$400000.0}
\expandafter\def\csname val@rev.desc.d3.uiv.p2.T.catalog-pooled.boot-resid.value\endcsname{-0.008}
\expandafter\def\csname cnt@rev.desc.d3.uiv.p2.T.catalog-pooled.boot-resid.value\endcsname{0}
\expandafter\def\csname pct@rev.desc.d3.uiv.p2.T.catalog-pooled.boot-resid.value\endcsname{$-$0.8}
\expandafter\def\csname ppt@rev.desc.d3.uiv.p2.T.catalog-pooled.boot-resid.value\endcsname{$-$0.8}
\expandafter\def\csname val@rev.desc.d3.uiv.p2.T.catalog-pooled.boot-share.ci95.0\endcsname{0.826}
\expandafter\def\csname cnt@rev.desc.d3.uiv.p2.T.catalog-pooled.boot-share.ci95.0\endcsname{1}
\expandafter\def\csname pct@rev.desc.d3.uiv.p2.T.catalog-pooled.boot-share.ci95.0\endcsname{82.6}
\expandafter\def\csname ppt@rev.desc.d3.uiv.p2.T.catalog-pooled.boot-share.ci95.0\endcsname{$+$82.6}
\expandafter\def\csname val@rev.desc.d3.uiv.p2.T.catalog-pooled.boot-share.ci95.1\endcsname{0.944}
\expandafter\def\csname cnt@rev.desc.d3.uiv.p2.T.catalog-pooled.boot-share.ci95.1\endcsname{1}
\expandafter\def\csname pct@rev.desc.d3.uiv.p2.T.catalog-pooled.boot-share.ci95.1\endcsname{94.4}
\expandafter\def\csname ppt@rev.desc.d3.uiv.p2.T.catalog-pooled.boot-share.ci95.1\endcsname{$+$94.4}
\expandafter\def\csname val@rev.desc.d3.uiv.p2.T.catalog-pooled.boot-share.n-apps\endcsname{24}
\expandafter\def\csname cnt@rev.desc.d3.uiv.p2.T.catalog-pooled.boot-share.n-apps\endcsname{24}
\expandafter\def\csname pct@rev.desc.d3.uiv.p2.T.catalog-pooled.boot-share.n-apps\endcsname{2400.0}
\expandafter\def\csname ppt@rev.desc.d3.uiv.p2.T.catalog-pooled.boot-share.n-apps\endcsname{$+$2400.0}
\expandafter\def\csname val@rev.desc.d3.uiv.p2.T.catalog-pooled.boot-share.n-boot-valid\endcsname{4000}
\expandafter\def\csname cnt@rev.desc.d3.uiv.p2.T.catalog-pooled.boot-share.n-boot-valid\endcsname{4{,}000}
\expandafter\def\csname pct@rev.desc.d3.uiv.p2.T.catalog-pooled.boot-share.n-boot-valid\endcsname{400000.0}
\expandafter\def\csname ppt@rev.desc.d3.uiv.p2.T.catalog-pooled.boot-share.n-boot-valid\endcsname{$+$400000.0}
\expandafter\def\csname val@rev.desc.d3.uiv.p2.T.catalog-pooled.boot-share.value\endcsname{0.889}
\expandafter\def\csname cnt@rev.desc.d3.uiv.p2.T.catalog-pooled.boot-share.value\endcsname{1}
\expandafter\def\csname pct@rev.desc.d3.uiv.p2.T.catalog-pooled.boot-share.value\endcsname{88.9}
\expandafter\def\csname ppt@rev.desc.d3.uiv.p2.T.catalog-pooled.boot-share.value\endcsname{$+$88.9}
\expandafter\def\csname val@rev.desc.d3.uiv.p2.T.catalog-pooled.gap\endcsname{0.199}
\expandafter\def\csname cnt@rev.desc.d3.uiv.p2.T.catalog-pooled.gap\endcsname{0}
\expandafter\def\csname pct@rev.desc.d3.uiv.p2.T.catalog-pooled.gap\endcsname{19.9}
\expandafter\def\csname ppt@rev.desc.d3.uiv.p2.T.catalog-pooled.gap\endcsname{$+$19.9}
\expandafter\def\csname val@rev.desc.d3.uiv.p2.T.catalog-pooled.n\endcsname{576}
\expandafter\def\csname cnt@rev.desc.d3.uiv.p2.T.catalog-pooled.n\endcsname{576}
\expandafter\def\csname pct@rev.desc.d3.uiv.p2.T.catalog-pooled.n\endcsname{57600.0}
\expandafter\def\csname ppt@rev.desc.d3.uiv.p2.T.catalog-pooled.n\endcsname{$+$57600.0}
\expandafter\def\csname val@rev.desc.d3.uiv.p2.T.catalog-pooled.reproduces-frozen\endcsname{1}
\expandafter\def\csname cnt@rev.desc.d3.uiv.p2.T.catalog-pooled.reproduces-frozen\endcsname{1}
\expandafter\def\csname pct@rev.desc.d3.uiv.p2.T.catalog-pooled.reproduces-frozen\endcsname{100.0}
\expandafter\def\csname ppt@rev.desc.d3.uiv.p2.T.catalog-pooled.reproduces-frozen\endcsname{$+$100.0}
\expandafter\def\csname val@rev.desc.d3.uiv.p2.T.catalog-pooled.resid-agree-stable\endcsname{-0.008}
\expandafter\def\csname cnt@rev.desc.d3.uiv.p2.T.catalog-pooled.resid-agree-stable\endcsname{0}
\expandafter\def\csname pct@rev.desc.d3.uiv.p2.T.catalog-pooled.resid-agree-stable\endcsname{$-$0.8}
\expandafter\def\csname ppt@rev.desc.d3.uiv.p2.T.catalog-pooled.resid-agree-stable\endcsname{$-$0.8}
\expandafter\def\csname val@rev.desc.d3.uiv.p2.T.catalog-pooled.resid-disagree\endcsname{0.191}
\expandafter\def\csname cnt@rev.desc.d3.uiv.p2.T.catalog-pooled.resid-disagree\endcsname{0}
\expandafter\def\csname pct@rev.desc.d3.uiv.p2.T.catalog-pooled.resid-disagree\endcsname{19.1}
\expandafter\def\csname ppt@rev.desc.d3.uiv.p2.T.catalog-pooled.resid-disagree\endcsname{$+$19.1}
\expandafter\def\csname val@rev.desc.d3.uiv.p2.T.catalog-pooled.share-agree-stable\endcsname{0.889}
\expandafter\def\csname cnt@rev.desc.d3.uiv.p2.T.catalog-pooled.share-agree-stable\endcsname{1}
\expandafter\def\csname pct@rev.desc.d3.uiv.p2.T.catalog-pooled.share-agree-stable\endcsname{88.9}
\expandafter\def\csname ppt@rev.desc.d3.uiv.p2.T.catalog-pooled.share-agree-stable\endcsname{$+$88.9}
\expandafter\def\csname val@rev.desc.d3.uiv.p2.T.center-d1.N-agree-stable\endcsname{128}
\expandafter\def\csname cnt@rev.desc.d3.uiv.p2.T.center-d1.N-agree-stable\endcsname{128}
\expandafter\def\csname pct@rev.desc.d3.uiv.p2.T.center-d1.N-agree-stable\endcsname{12800.0}
\expandafter\def\csname ppt@rev.desc.d3.uiv.p2.T.center-d1.N-agree-stable\endcsname{$+$12800.0}
\expandafter\def\csname val@rev.desc.d3.uiv.p2.T.center-d1.N-disagree\endcsname{11}
\expandafter\def\csname cnt@rev.desc.d3.uiv.p2.T.center-d1.N-disagree\endcsname{11}
\expandafter\def\csname pct@rev.desc.d3.uiv.p2.T.center-d1.N-disagree\endcsname{1100.0}
\expandafter\def\csname ppt@rev.desc.d3.uiv.p2.T.center-d1.N-disagree\endcsname{$+$1100.0}
\expandafter\def\csname val@rev.desc.d3.uiv.p2.T.center-d1.boot-resid.ci95.0\endcsname{-0.024}
\expandafter\def\csname cnt@rev.desc.d3.uiv.p2.T.center-d1.boot-resid.ci95.0\endcsname{0}
\expandafter\def\csname pct@rev.desc.d3.uiv.p2.T.center-d1.boot-resid.ci95.0\endcsname{$-$2.4}
\expandafter\def\csname ppt@rev.desc.d3.uiv.p2.T.center-d1.boot-resid.ci95.0\endcsname{$-$2.4}
\expandafter\def\csname val@rev.desc.d3.uiv.p2.T.center-d1.boot-resid.ci95.1\endcsname{0.000}
\expandafter\def\csname cnt@rev.desc.d3.uiv.p2.T.center-d1.boot-resid.ci95.1\endcsname{0}
\expandafter\def\csname pct@rev.desc.d3.uiv.p2.T.center-d1.boot-resid.ci95.1\endcsname{0.0}
\expandafter\def\csname ppt@rev.desc.d3.uiv.p2.T.center-d1.boot-resid.ci95.1\endcsname{0.0}
\expandafter\def\csname val@rev.desc.d3.uiv.p2.T.center-d1.boot-resid.n-apps\endcsname{24}
\expandafter\def\csname cnt@rev.desc.d3.uiv.p2.T.center-d1.boot-resid.n-apps\endcsname{24}
\expandafter\def\csname pct@rev.desc.d3.uiv.p2.T.center-d1.boot-resid.n-apps\endcsname{2400.0}
\expandafter\def\csname ppt@rev.desc.d3.uiv.p2.T.center-d1.boot-resid.n-apps\endcsname{$+$2400.0}
\expandafter\def\csname val@rev.desc.d3.uiv.p2.T.center-d1.boot-resid.n-boot-valid\endcsname{4000}
\expandafter\def\csname cnt@rev.desc.d3.uiv.p2.T.center-d1.boot-resid.n-boot-valid\endcsname{4{,}000}
\expandafter\def\csname pct@rev.desc.d3.uiv.p2.T.center-d1.boot-resid.n-boot-valid\endcsname{400000.0}
\expandafter\def\csname ppt@rev.desc.d3.uiv.p2.T.center-d1.boot-resid.n-boot-valid\endcsname{$+$400000.0}
\expandafter\def\csname val@rev.desc.d3.uiv.p2.T.center-d1.boot-resid.value\endcsname{-0.008}
\expandafter\def\csname cnt@rev.desc.d3.uiv.p2.T.center-d1.boot-resid.value\endcsname{0}
\expandafter\def\csname pct@rev.desc.d3.uiv.p2.T.center-d1.boot-resid.value\endcsname{$-$0.8}
\expandafter\def\csname ppt@rev.desc.d3.uiv.p2.T.center-d1.boot-resid.value\endcsname{$-$0.8}
\expandafter\def\csname val@rev.desc.d3.uiv.p2.T.center-d1.boot-share.ci95.0\endcsname{0.812}
\expandafter\def\csname cnt@rev.desc.d3.uiv.p2.T.center-d1.boot-share.ci95.0\endcsname{1}
\expandafter\def\csname pct@rev.desc.d3.uiv.p2.T.center-d1.boot-share.ci95.0\endcsname{81.2}
\expandafter\def\csname ppt@rev.desc.d3.uiv.p2.T.center-d1.boot-share.ci95.0\endcsname{$+$81.2}
\expandafter\def\csname val@rev.desc.d3.uiv.p2.T.center-d1.boot-share.ci95.1\endcsname{0.958}
\expandafter\def\csname cnt@rev.desc.d3.uiv.p2.T.center-d1.boot-share.ci95.1\endcsname{1}
\expandafter\def\csname pct@rev.desc.d3.uiv.p2.T.center-d1.boot-share.ci95.1\endcsname{95.8}
\expandafter\def\csname ppt@rev.desc.d3.uiv.p2.T.center-d1.boot-share.ci95.1\endcsname{$+$95.8}
\expandafter\def\csname val@rev.desc.d3.uiv.p2.T.center-d1.boot-share.n-apps\endcsname{24}
\expandafter\def\csname cnt@rev.desc.d3.uiv.p2.T.center-d1.boot-share.n-apps\endcsname{24}
\expandafter\def\csname pct@rev.desc.d3.uiv.p2.T.center-d1.boot-share.n-apps\endcsname{2400.0}
\expandafter\def\csname ppt@rev.desc.d3.uiv.p2.T.center-d1.boot-share.n-apps\endcsname{$+$2400.0}
\expandafter\def\csname val@rev.desc.d3.uiv.p2.T.center-d1.boot-share.n-boot-valid\endcsname{4000}
\expandafter\def\csname cnt@rev.desc.d3.uiv.p2.T.center-d1.boot-share.n-boot-valid\endcsname{4{,}000}
\expandafter\def\csname pct@rev.desc.d3.uiv.p2.T.center-d1.boot-share.n-boot-valid\endcsname{400000.0}
\expandafter\def\csname ppt@rev.desc.d3.uiv.p2.T.center-d1.boot-share.n-boot-valid\endcsname{$+$400000.0}
\expandafter\def\csname val@rev.desc.d3.uiv.p2.T.center-d1.boot-share.value\endcsname{0.889}
\expandafter\def\csname cnt@rev.desc.d3.uiv.p2.T.center-d1.boot-share.value\endcsname{1}
\expandafter\def\csname pct@rev.desc.d3.uiv.p2.T.center-d1.boot-share.value\endcsname{88.9}
\expandafter\def\csname ppt@rev.desc.d3.uiv.p2.T.center-d1.boot-share.value\endcsname{$+$88.9}
\expandafter\def\csname val@rev.desc.d3.uiv.p2.T.center-d1.n\endcsname{144}
\expandafter\def\csname cnt@rev.desc.d3.uiv.p2.T.center-d1.n\endcsname{144}
\expandafter\def\csname pct@rev.desc.d3.uiv.p2.T.center-d1.n\endcsname{14400.0}
\expandafter\def\csname ppt@rev.desc.d3.uiv.p2.T.center-d1.n\endcsname{$+$14400.0}
\expandafter\def\csname val@rev.desc.d3.uiv.p2.T.center-d1.reproduces-frozen\endcsname{1}
\expandafter\def\csname cnt@rev.desc.d3.uiv.p2.T.center-d1.reproduces-frozen\endcsname{1}
\expandafter\def\csname pct@rev.desc.d3.uiv.p2.T.center-d1.reproduces-frozen\endcsname{100.0}
\expandafter\def\csname ppt@rev.desc.d3.uiv.p2.T.center-d1.reproduces-frozen\endcsname{$+$100.0}
\expandafter\def\csname val@rev.desc.d3.uiv.p2.T.center-d1.resid-agree-stable\endcsname{-0.008}
\expandafter\def\csname cnt@rev.desc.d3.uiv.p2.T.center-d1.resid-agree-stable\endcsname{0}
\expandafter\def\csname pct@rev.desc.d3.uiv.p2.T.center-d1.resid-agree-stable\endcsname{$-$0.8}
\expandafter\def\csname ppt@rev.desc.d3.uiv.p2.T.center-d1.resid-agree-stable\endcsname{$-$0.8}
\expandafter\def\csname val@rev.desc.d3.uiv.p2.T.center-d1.resid-disagree\endcsname{0.000}
\expandafter\def\csname cnt@rev.desc.d3.uiv.p2.T.center-d1.resid-disagree\endcsname{0}
\expandafter\def\csname pct@rev.desc.d3.uiv.p2.T.center-d1.resid-disagree\endcsname{0.0}
\expandafter\def\csname ppt@rev.desc.d3.uiv.p2.T.center-d1.resid-disagree\endcsname{0.0}
\expandafter\def\csname val@rev.desc.d3.uiv.p2.T.center-d1.share-agree-stable\endcsname{0.889}
\expandafter\def\csname cnt@rev.desc.d3.uiv.p2.T.center-d1.share-agree-stable\endcsname{1}
\expandafter\def\csname pct@rev.desc.d3.uiv.p2.T.center-d1.share-agree-stable\endcsname{88.9}
\expandafter\def\csname ppt@rev.desc.d3.uiv.p2.T.center-d1.share-agree-stable\endcsname{$+$88.9}
\expandafter\def\csname val@rev.desc.d3.uiv.p2.T.context-d1.N-agree-stable\endcsname{131}
\expandafter\def\csname cnt@rev.desc.d3.uiv.p2.T.context-d1.N-agree-stable\endcsname{131}
\expandafter\def\csname pct@rev.desc.d3.uiv.p2.T.context-d1.N-agree-stable\endcsname{13100.0}
\expandafter\def\csname ppt@rev.desc.d3.uiv.p2.T.context-d1.N-agree-stable\endcsname{$+$13100.0}
\expandafter\def\csname val@rev.desc.d3.uiv.p2.T.context-d1.N-disagree\endcsname{10}
\expandafter\def\csname cnt@rev.desc.d3.uiv.p2.T.context-d1.N-disagree\endcsname{10}
\expandafter\def\csname pct@rev.desc.d3.uiv.p2.T.context-d1.N-disagree\endcsname{1000.0}
\expandafter\def\csname ppt@rev.desc.d3.uiv.p2.T.context-d1.N-disagree\endcsname{$+$1000.0}
\expandafter\def\csname val@rev.desc.d3.uiv.p2.T.context-d1.boot-resid.ci95.0\endcsname{-0.023}
\expandafter\def\csname cnt@rev.desc.d3.uiv.p2.T.context-d1.boot-resid.ci95.0\endcsname{0}
\expandafter\def\csname pct@rev.desc.d3.uiv.p2.T.context-d1.boot-resid.ci95.0\endcsname{$-$2.3}
\expandafter\def\csname ppt@rev.desc.d3.uiv.p2.T.context-d1.boot-resid.ci95.0\endcsname{$-$2.3}
\expandafter\def\csname val@rev.desc.d3.uiv.p2.T.context-d1.boot-resid.ci95.1\endcsname{0.023}
\expandafter\def\csname cnt@rev.desc.d3.uiv.p2.T.context-d1.boot-resid.ci95.1\endcsname{0}
\expandafter\def\csname pct@rev.desc.d3.uiv.p2.T.context-d1.boot-resid.ci95.1\endcsname{2.3}
\expandafter\def\csname ppt@rev.desc.d3.uiv.p2.T.context-d1.boot-resid.ci95.1\endcsname{$+$2.3}
\expandafter\def\csname val@rev.desc.d3.uiv.p2.T.context-d1.boot-resid.n-apps\endcsname{24}
\expandafter\def\csname cnt@rev.desc.d3.uiv.p2.T.context-d1.boot-resid.n-apps\endcsname{24}
\expandafter\def\csname pct@rev.desc.d3.uiv.p2.T.context-d1.boot-resid.n-apps\endcsname{2400.0}
\expandafter\def\csname ppt@rev.desc.d3.uiv.p2.T.context-d1.boot-resid.n-apps\endcsname{$+$2400.0}
\expandafter\def\csname val@rev.desc.d3.uiv.p2.T.context-d1.boot-resid.n-boot-valid\endcsname{4000}
\expandafter\def\csname cnt@rev.desc.d3.uiv.p2.T.context-d1.boot-resid.n-boot-valid\endcsname{4{,}000}
\expandafter\def\csname pct@rev.desc.d3.uiv.p2.T.context-d1.boot-resid.n-boot-valid\endcsname{400000.0}
\expandafter\def\csname ppt@rev.desc.d3.uiv.p2.T.context-d1.boot-resid.n-boot-valid\endcsname{$+$400000.0}
\expandafter\def\csname val@rev.desc.d3.uiv.p2.T.context-d1.boot-resid.value\endcsname{0.000}
\expandafter\def\csname cnt@rev.desc.d3.uiv.p2.T.context-d1.boot-resid.value\endcsname{0}
\expandafter\def\csname pct@rev.desc.d3.uiv.p2.T.context-d1.boot-resid.value\endcsname{0.0}
\expandafter\def\csname ppt@rev.desc.d3.uiv.p2.T.context-d1.boot-resid.value\endcsname{0.0}
\expandafter\def\csname val@rev.desc.d3.uiv.p2.T.context-d1.boot-share.ci95.0\endcsname{0.854}
\expandafter\def\csname cnt@rev.desc.d3.uiv.p2.T.context-d1.boot-share.ci95.0\endcsname{1}
\expandafter\def\csname pct@rev.desc.d3.uiv.p2.T.context-d1.boot-share.ci95.0\endcsname{85.4}
\expandafter\def\csname ppt@rev.desc.d3.uiv.p2.T.context-d1.boot-share.ci95.0\endcsname{$+$85.4}
\expandafter\def\csname val@rev.desc.d3.uiv.p2.T.context-d1.boot-share.ci95.1\endcsname{0.958}
\expandafter\def\csname cnt@rev.desc.d3.uiv.p2.T.context-d1.boot-share.ci95.1\endcsname{1}
\expandafter\def\csname pct@rev.desc.d3.uiv.p2.T.context-d1.boot-share.ci95.1\endcsname{95.8}
\expandafter\def\csname ppt@rev.desc.d3.uiv.p2.T.context-d1.boot-share.ci95.1\endcsname{$+$95.8}
\expandafter\def\csname val@rev.desc.d3.uiv.p2.T.context-d1.boot-share.n-apps\endcsname{24}
\expandafter\def\csname cnt@rev.desc.d3.uiv.p2.T.context-d1.boot-share.n-apps\endcsname{24}
\expandafter\def\csname pct@rev.desc.d3.uiv.p2.T.context-d1.boot-share.n-apps\endcsname{2400.0}
\expandafter\def\csname ppt@rev.desc.d3.uiv.p2.T.context-d1.boot-share.n-apps\endcsname{$+$2400.0}
\expandafter\def\csname val@rev.desc.d3.uiv.p2.T.context-d1.boot-share.n-boot-valid\endcsname{4000}
\expandafter\def\csname cnt@rev.desc.d3.uiv.p2.T.context-d1.boot-share.n-boot-valid\endcsname{4{,}000}
\expandafter\def\csname pct@rev.desc.d3.uiv.p2.T.context-d1.boot-share.n-boot-valid\endcsname{400000.0}
\expandafter\def\csname ppt@rev.desc.d3.uiv.p2.T.context-d1.boot-share.n-boot-valid\endcsname{$+$400000.0}
\expandafter\def\csname val@rev.desc.d3.uiv.p2.T.context-d1.boot-share.value\endcsname{0.910}
\expandafter\def\csname cnt@rev.desc.d3.uiv.p2.T.context-d1.boot-share.value\endcsname{1}
\expandafter\def\csname pct@rev.desc.d3.uiv.p2.T.context-d1.boot-share.value\endcsname{91.0}
\expandafter\def\csname ppt@rev.desc.d3.uiv.p2.T.context-d1.boot-share.value\endcsname{$+$91.0}
\expandafter\def\csname val@rev.desc.d3.uiv.p2.T.context-d1.n\endcsname{144}
\expandafter\def\csname cnt@rev.desc.d3.uiv.p2.T.context-d1.n\endcsname{144}
\expandafter\def\csname pct@rev.desc.d3.uiv.p2.T.context-d1.n\endcsname{14400.0}
\expandafter\def\csname ppt@rev.desc.d3.uiv.p2.T.context-d1.n\endcsname{$+$14400.0}
\expandafter\def\csname val@rev.desc.d3.uiv.p2.T.context-d1.reproduces-frozen\endcsname{1}
\expandafter\def\csname cnt@rev.desc.d3.uiv.p2.T.context-d1.reproduces-frozen\endcsname{1}
\expandafter\def\csname pct@rev.desc.d3.uiv.p2.T.context-d1.reproduces-frozen\endcsname{100.0}
\expandafter\def\csname ppt@rev.desc.d3.uiv.p2.T.context-d1.reproduces-frozen\endcsname{$+$100.0}
\expandafter\def\csname val@rev.desc.d3.uiv.p2.T.context-d1.resid-agree-stable\endcsname{0.000}
\expandafter\def\csname cnt@rev.desc.d3.uiv.p2.T.context-d1.resid-agree-stable\endcsname{0}
\expandafter\def\csname pct@rev.desc.d3.uiv.p2.T.context-d1.resid-agree-stable\endcsname{0.0}
\expandafter\def\csname ppt@rev.desc.d3.uiv.p2.T.context-d1.resid-agree-stable\endcsname{0.0}
\expandafter\def\csname val@rev.desc.d3.uiv.p2.T.context-d1.resid-disagree\endcsname{0.400}
\expandafter\def\csname cnt@rev.desc.d3.uiv.p2.T.context-d1.resid-disagree\endcsname{0}
\expandafter\def\csname pct@rev.desc.d3.uiv.p2.T.context-d1.resid-disagree\endcsname{40.0}
\expandafter\def\csname ppt@rev.desc.d3.uiv.p2.T.context-d1.resid-disagree\endcsname{$+$40.0}
\expandafter\def\csname val@rev.desc.d3.uiv.p2.T.context-d1.share-agree-stable\endcsname{0.910}
\expandafter\def\csname cnt@rev.desc.d3.uiv.p2.T.context-d1.share-agree-stable\endcsname{1}
\expandafter\def\csname pct@rev.desc.d3.uiv.p2.T.context-d1.share-agree-stable\endcsname{91.0}
\expandafter\def\csname ppt@rev.desc.d3.uiv.p2.T.context-d1.share-agree-stable\endcsname{$+$91.0}
\expandafter\def\csname val@rev.desc.d3.uiv.p2.T.dual-d2.N-agree-stable\endcsname{129}
\expandafter\def\csname cnt@rev.desc.d3.uiv.p2.T.dual-d2.N-agree-stable\endcsname{129}
\expandafter\def\csname pct@rev.desc.d3.uiv.p2.T.dual-d2.N-agree-stable\endcsname{12900.0}
\expandafter\def\csname ppt@rev.desc.d3.uiv.p2.T.dual-d2.N-agree-stable\endcsname{$+$12900.0}
\expandafter\def\csname val@rev.desc.d3.uiv.p2.T.dual-d2.N-disagree\endcsname{10}
\expandafter\def\csname cnt@rev.desc.d3.uiv.p2.T.dual-d2.N-disagree\endcsname{10}
\expandafter\def\csname pct@rev.desc.d3.uiv.p2.T.dual-d2.N-disagree\endcsname{1000.0}
\expandafter\def\csname ppt@rev.desc.d3.uiv.p2.T.dual-d2.N-disagree\endcsname{$+$1000.0}
\expandafter\def\csname val@rev.desc.d3.uiv.p2.T.dual-d2.boot-resid.ci95.0\endcsname{-0.054}
\expandafter\def\csname cnt@rev.desc.d3.uiv.p2.T.dual-d2.boot-resid.ci95.0\endcsname{0}
\expandafter\def\csname pct@rev.desc.d3.uiv.p2.T.dual-d2.boot-resid.ci95.0\endcsname{$-$5.4}
\expandafter\def\csname ppt@rev.desc.d3.uiv.p2.T.dual-d2.boot-resid.ci95.0\endcsname{$-$5.4}
\expandafter\def\csname val@rev.desc.d3.uiv.p2.T.dual-d2.boot-resid.ci95.1\endcsname{0.039}
\expandafter\def\csname cnt@rev.desc.d3.uiv.p2.T.dual-d2.boot-resid.ci95.1\endcsname{0}
\expandafter\def\csname pct@rev.desc.d3.uiv.p2.T.dual-d2.boot-resid.ci95.1\endcsname{3.9}
\expandafter\def\csname ppt@rev.desc.d3.uiv.p2.T.dual-d2.boot-resid.ci95.1\endcsname{$+$3.9}
\expandafter\def\csname val@rev.desc.d3.uiv.p2.T.dual-d2.boot-resid.n-apps\endcsname{24}
\expandafter\def\csname cnt@rev.desc.d3.uiv.p2.T.dual-d2.boot-resid.n-apps\endcsname{24}
\expandafter\def\csname pct@rev.desc.d3.uiv.p2.T.dual-d2.boot-resid.n-apps\endcsname{2400.0}
\expandafter\def\csname ppt@rev.desc.d3.uiv.p2.T.dual-d2.boot-resid.n-apps\endcsname{$+$2400.0}
\expandafter\def\csname val@rev.desc.d3.uiv.p2.T.dual-d2.boot-resid.n-boot-valid\endcsname{4000}
\expandafter\def\csname cnt@rev.desc.d3.uiv.p2.T.dual-d2.boot-resid.n-boot-valid\endcsname{4{,}000}
\expandafter\def\csname pct@rev.desc.d3.uiv.p2.T.dual-d2.boot-resid.n-boot-valid\endcsname{400000.0}
\expandafter\def\csname ppt@rev.desc.d3.uiv.p2.T.dual-d2.boot-resid.n-boot-valid\endcsname{$+$400000.0}
\expandafter\def\csname val@rev.desc.d3.uiv.p2.T.dual-d2.boot-resid.value\endcsname{-0.008}
\expandafter\def\csname cnt@rev.desc.d3.uiv.p2.T.dual-d2.boot-resid.value\endcsname{0}
\expandafter\def\csname pct@rev.desc.d3.uiv.p2.T.dual-d2.boot-resid.value\endcsname{$-$0.8}
\expandafter\def\csname ppt@rev.desc.d3.uiv.p2.T.dual-d2.boot-resid.value\endcsname{$-$0.8}
\expandafter\def\csname val@rev.desc.d3.uiv.p2.T.dual-d2.boot-share.ci95.0\endcsname{0.833}
\expandafter\def\csname cnt@rev.desc.d3.uiv.p2.T.dual-d2.boot-share.ci95.0\endcsname{1}
\expandafter\def\csname pct@rev.desc.d3.uiv.p2.T.dual-d2.boot-share.ci95.0\endcsname{83.3}
\expandafter\def\csname ppt@rev.desc.d3.uiv.p2.T.dual-d2.boot-share.ci95.0\endcsname{$+$83.3}
\expandafter\def\csname val@rev.desc.d3.uiv.p2.T.dual-d2.boot-share.ci95.1\endcsname{0.951}
\expandafter\def\csname cnt@rev.desc.d3.uiv.p2.T.dual-d2.boot-share.ci95.1\endcsname{1}
\expandafter\def\csname pct@rev.desc.d3.uiv.p2.T.dual-d2.boot-share.ci95.1\endcsname{95.1}
\expandafter\def\csname ppt@rev.desc.d3.uiv.p2.T.dual-d2.boot-share.ci95.1\endcsname{$+$95.1}
\expandafter\def\csname val@rev.desc.d3.uiv.p2.T.dual-d2.boot-share.n-apps\endcsname{24}
\expandafter\def\csname cnt@rev.desc.d3.uiv.p2.T.dual-d2.boot-share.n-apps\endcsname{24}
\expandafter\def\csname pct@rev.desc.d3.uiv.p2.T.dual-d2.boot-share.n-apps\endcsname{2400.0}
\expandafter\def\csname ppt@rev.desc.d3.uiv.p2.T.dual-d2.boot-share.n-apps\endcsname{$+$2400.0}
\expandafter\def\csname val@rev.desc.d3.uiv.p2.T.dual-d2.boot-share.n-boot-valid\endcsname{4000}
\expandafter\def\csname cnt@rev.desc.d3.uiv.p2.T.dual-d2.boot-share.n-boot-valid\endcsname{4{,}000}
\expandafter\def\csname pct@rev.desc.d3.uiv.p2.T.dual-d2.boot-share.n-boot-valid\endcsname{400000.0}
\expandafter\def\csname ppt@rev.desc.d3.uiv.p2.T.dual-d2.boot-share.n-boot-valid\endcsname{$+$400000.0}
\expandafter\def\csname val@rev.desc.d3.uiv.p2.T.dual-d2.boot-share.value\endcsname{0.896}
\expandafter\def\csname cnt@rev.desc.d3.uiv.p2.T.dual-d2.boot-share.value\endcsname{1}
\expandafter\def\csname pct@rev.desc.d3.uiv.p2.T.dual-d2.boot-share.value\endcsname{89.6}
\expandafter\def\csname ppt@rev.desc.d3.uiv.p2.T.dual-d2.boot-share.value\endcsname{$+$89.6}
\expandafter\def\csname val@rev.desc.d3.uiv.p2.T.dual-d2.n\endcsname{144}
\expandafter\def\csname cnt@rev.desc.d3.uiv.p2.T.dual-d2.n\endcsname{144}
\expandafter\def\csname pct@rev.desc.d3.uiv.p2.T.dual-d2.n\endcsname{14400.0}
\expandafter\def\csname ppt@rev.desc.d3.uiv.p2.T.dual-d2.n\endcsname{$+$14400.0}
\expandafter\def\csname val@rev.desc.d3.uiv.p2.T.dual-d2.reproduces-frozen\endcsname{1}
\expandafter\def\csname cnt@rev.desc.d3.uiv.p2.T.dual-d2.reproduces-frozen\endcsname{1}
\expandafter\def\csname pct@rev.desc.d3.uiv.p2.T.dual-d2.reproduces-frozen\endcsname{100.0}
\expandafter\def\csname ppt@rev.desc.d3.uiv.p2.T.dual-d2.reproduces-frozen\endcsname{$+$100.0}
\expandafter\def\csname val@rev.desc.d3.uiv.p2.T.dual-d2.resid-agree-stable\endcsname{-0.008}
\expandafter\def\csname cnt@rev.desc.d3.uiv.p2.T.dual-d2.resid-agree-stable\endcsname{0}
\expandafter\def\csname pct@rev.desc.d3.uiv.p2.T.dual-d2.resid-agree-stable\endcsname{$-$0.8}
\expandafter\def\csname ppt@rev.desc.d3.uiv.p2.T.dual-d2.resid-agree-stable\endcsname{$-$0.8}
\expandafter\def\csname val@rev.desc.d3.uiv.p2.T.dual-d2.resid-disagree\endcsname{-0.100}
\expandafter\def\csname cnt@rev.desc.d3.uiv.p2.T.dual-d2.resid-disagree\endcsname{0}
\expandafter\def\csname pct@rev.desc.d3.uiv.p2.T.dual-d2.resid-disagree\endcsname{$-$10.0}
\expandafter\def\csname ppt@rev.desc.d3.uiv.p2.T.dual-d2.resid-disagree\endcsname{$-$10.0}
\expandafter\def\csname val@rev.desc.d3.uiv.p2.T.dual-d2.share-agree-stable\endcsname{0.896}
\expandafter\def\csname cnt@rev.desc.d3.uiv.p2.T.dual-d2.share-agree-stable\endcsname{1}
\expandafter\def\csname pct@rev.desc.d3.uiv.p2.T.dual-d2.share-agree-stable\endcsname{89.6}
\expandafter\def\csname ppt@rev.desc.d3.uiv.p2.T.dual-d2.share-agree-stable\endcsname{$+$89.6}
\expandafter\def\csname val@rev.desc.d3.uiv.p2.T.grid2-d3.N-agree-stable\endcsname{124}
\expandafter\def\csname cnt@rev.desc.d3.uiv.p2.T.grid2-d3.N-agree-stable\endcsname{124}
\expandafter\def\csname pct@rev.desc.d3.uiv.p2.T.grid2-d3.N-agree-stable\endcsname{12400.0}
\expandafter\def\csname ppt@rev.desc.d3.uiv.p2.T.grid2-d3.N-agree-stable\endcsname{$+$12400.0}
\expandafter\def\csname val@rev.desc.d3.uiv.p2.T.grid2-d3.N-disagree\endcsname{16}
\expandafter\def\csname cnt@rev.desc.d3.uiv.p2.T.grid2-d3.N-disagree\endcsname{16}
\expandafter\def\csname pct@rev.desc.d3.uiv.p2.T.grid2-d3.N-disagree\endcsname{1600.0}
\expandafter\def\csname ppt@rev.desc.d3.uiv.p2.T.grid2-d3.N-disagree\endcsname{$+$1600.0}
\expandafter\def\csname val@rev.desc.d3.uiv.p2.T.grid2-d3.boot-resid.ci95.0\endcsname{-0.042}
\expandafter\def\csname cnt@rev.desc.d3.uiv.p2.T.grid2-d3.boot-resid.ci95.0\endcsname{0}
\expandafter\def\csname pct@rev.desc.d3.uiv.p2.T.grid2-d3.boot-resid.ci95.0\endcsname{$-$4.2}
\expandafter\def\csname ppt@rev.desc.d3.uiv.p2.T.grid2-d3.boot-resid.ci95.0\endcsname{$-$4.2}
\expandafter\def\csname val@rev.desc.d3.uiv.p2.T.grid2-d3.boot-resid.ci95.1\endcsname{0.000}
\expandafter\def\csname cnt@rev.desc.d3.uiv.p2.T.grid2-d3.boot-resid.ci95.1\endcsname{0}
\expandafter\def\csname pct@rev.desc.d3.uiv.p2.T.grid2-d3.boot-resid.ci95.1\endcsname{0.0}
\expandafter\def\csname ppt@rev.desc.d3.uiv.p2.T.grid2-d3.boot-resid.ci95.1\endcsname{0.0}
\expandafter\def\csname val@rev.desc.d3.uiv.p2.T.grid2-d3.boot-resid.n-apps\endcsname{24}
\expandafter\def\csname cnt@rev.desc.d3.uiv.p2.T.grid2-d3.boot-resid.n-apps\endcsname{24}
\expandafter\def\csname pct@rev.desc.d3.uiv.p2.T.grid2-d3.boot-resid.n-apps\endcsname{2400.0}
\expandafter\def\csname ppt@rev.desc.d3.uiv.p2.T.grid2-d3.boot-resid.n-apps\endcsname{$+$2400.0}
\expandafter\def\csname val@rev.desc.d3.uiv.p2.T.grid2-d3.boot-resid.n-boot-valid\endcsname{4000}
\expandafter\def\csname cnt@rev.desc.d3.uiv.p2.T.grid2-d3.boot-resid.n-boot-valid\endcsname{4{,}000}
\expandafter\def\csname pct@rev.desc.d3.uiv.p2.T.grid2-d3.boot-resid.n-boot-valid\endcsname{400000.0}
\expandafter\def\csname ppt@rev.desc.d3.uiv.p2.T.grid2-d3.boot-resid.n-boot-valid\endcsname{$+$400000.0}
\expandafter\def\csname val@rev.desc.d3.uiv.p2.T.grid2-d3.boot-resid.value\endcsname{-0.016}
\expandafter\def\csname cnt@rev.desc.d3.uiv.p2.T.grid2-d3.boot-resid.value\endcsname{0}
\expandafter\def\csname pct@rev.desc.d3.uiv.p2.T.grid2-d3.boot-resid.value\endcsname{$-$1.6}
\expandafter\def\csname ppt@rev.desc.d3.uiv.p2.T.grid2-d3.boot-resid.value\endcsname{$-$1.6}
\expandafter\def\csname val@rev.desc.d3.uiv.p2.T.grid2-d3.boot-share.ci95.0\endcsname{0.785}
\expandafter\def\csname cnt@rev.desc.d3.uiv.p2.T.grid2-d3.boot-share.ci95.0\endcsname{1}
\expandafter\def\csname pct@rev.desc.d3.uiv.p2.T.grid2-d3.boot-share.ci95.0\endcsname{78.5}
\expandafter\def\csname ppt@rev.desc.d3.uiv.p2.T.grid2-d3.boot-share.ci95.0\endcsname{$+$78.5}
\expandafter\def\csname val@rev.desc.d3.uiv.p2.T.grid2-d3.boot-share.ci95.1\endcsname{0.924}
\expandafter\def\csname cnt@rev.desc.d3.uiv.p2.T.grid2-d3.boot-share.ci95.1\endcsname{1}
\expandafter\def\csname pct@rev.desc.d3.uiv.p2.T.grid2-d3.boot-share.ci95.1\endcsname{92.4}
\expandafter\def\csname ppt@rev.desc.d3.uiv.p2.T.grid2-d3.boot-share.ci95.1\endcsname{$+$92.4}
\expandafter\def\csname val@rev.desc.d3.uiv.p2.T.grid2-d3.boot-share.n-apps\endcsname{24}
\expandafter\def\csname cnt@rev.desc.d3.uiv.p2.T.grid2-d3.boot-share.n-apps\endcsname{24}
\expandafter\def\csname pct@rev.desc.d3.uiv.p2.T.grid2-d3.boot-share.n-apps\endcsname{2400.0}
\expandafter\def\csname ppt@rev.desc.d3.uiv.p2.T.grid2-d3.boot-share.n-apps\endcsname{$+$2400.0}
\expandafter\def\csname val@rev.desc.d3.uiv.p2.T.grid2-d3.boot-share.n-boot-valid\endcsname{4000}
\expandafter\def\csname cnt@rev.desc.d3.uiv.p2.T.grid2-d3.boot-share.n-boot-valid\endcsname{4{,}000}
\expandafter\def\csname pct@rev.desc.d3.uiv.p2.T.grid2-d3.boot-share.n-boot-valid\endcsname{400000.0}
\expandafter\def\csname ppt@rev.desc.d3.uiv.p2.T.grid2-d3.boot-share.n-boot-valid\endcsname{$+$400000.0}
\expandafter\def\csname val@rev.desc.d3.uiv.p2.T.grid2-d3.boot-share.value\endcsname{0.861}
\expandafter\def\csname cnt@rev.desc.d3.uiv.p2.T.grid2-d3.boot-share.value\endcsname{1}
\expandafter\def\csname pct@rev.desc.d3.uiv.p2.T.grid2-d3.boot-share.value\endcsname{86.1}
\expandafter\def\csname ppt@rev.desc.d3.uiv.p2.T.grid2-d3.boot-share.value\endcsname{$+$86.1}
\expandafter\def\csname val@rev.desc.d3.uiv.p2.T.grid2-d3.n\endcsname{144}
\expandafter\def\csname cnt@rev.desc.d3.uiv.p2.T.grid2-d3.n\endcsname{144}
\expandafter\def\csname pct@rev.desc.d3.uiv.p2.T.grid2-d3.n\endcsname{14400.0}
\expandafter\def\csname ppt@rev.desc.d3.uiv.p2.T.grid2-d3.n\endcsname{$+$14400.0}
\expandafter\def\csname val@rev.desc.d3.uiv.p2.T.grid2-d3.reproduces-frozen\endcsname{1}
\expandafter\def\csname cnt@rev.desc.d3.uiv.p2.T.grid2-d3.reproduces-frozen\endcsname{1}
\expandafter\def\csname pct@rev.desc.d3.uiv.p2.T.grid2-d3.reproduces-frozen\endcsname{100.0}
\expandafter\def\csname ppt@rev.desc.d3.uiv.p2.T.grid2-d3.reproduces-frozen\endcsname{$+$100.0}
\expandafter\def\csname val@rev.desc.d3.uiv.p2.T.grid2-d3.resid-agree-stable\endcsname{-0.016}
\expandafter\def\csname cnt@rev.desc.d3.uiv.p2.T.grid2-d3.resid-agree-stable\endcsname{0}
\expandafter\def\csname pct@rev.desc.d3.uiv.p2.T.grid2-d3.resid-agree-stable\endcsname{$-$1.6}
\expandafter\def\csname ppt@rev.desc.d3.uiv.p2.T.grid2-d3.resid-agree-stable\endcsname{$-$1.6}
\expandafter\def\csname val@rev.desc.d3.uiv.p2.T.grid2-d3.resid-disagree\endcsname{0.375}
\expandafter\def\csname cnt@rev.desc.d3.uiv.p2.T.grid2-d3.resid-disagree\endcsname{0}
\expandafter\def\csname pct@rev.desc.d3.uiv.p2.T.grid2-d3.resid-disagree\endcsname{37.5}
\expandafter\def\csname ppt@rev.desc.d3.uiv.p2.T.grid2-d3.resid-disagree\endcsname{$+$37.5}
\expandafter\def\csname val@rev.desc.d3.uiv.p2.T.grid2-d3.share-agree-stable\endcsname{0.861}
\expandafter\def\csname cnt@rev.desc.d3.uiv.p2.T.grid2-d3.share-agree-stable\endcsname{1}
\expandafter\def\csname pct@rev.desc.d3.uiv.p2.T.grid2-d3.share-agree-stable\endcsname{86.1}
\expandafter\def\csname ppt@rev.desc.d3.uiv.p2.T.grid2-d3.share-agree-stable\endcsname{$+$86.1}
\expandafter\def\csname val@rev.desc.d3.uiv.p2.T.zoomclick-d3.N-agree-stable\endcsname{126}
\expandafter\def\csname cnt@rev.desc.d3.uiv.p2.T.zoomclick-d3.N-agree-stable\endcsname{126}
\expandafter\def\csname pct@rev.desc.d3.uiv.p2.T.zoomclick-d3.N-agree-stable\endcsname{12600.0}
\expandafter\def\csname ppt@rev.desc.d3.uiv.p2.T.zoomclick-d3.N-agree-stable\endcsname{$+$12600.0}
\expandafter\def\csname val@rev.desc.d3.uiv.p2.T.zoomclick-d3.N-disagree\endcsname{14}
\expandafter\def\csname cnt@rev.desc.d3.uiv.p2.T.zoomclick-d3.N-disagree\endcsname{14}
\expandafter\def\csname pct@rev.desc.d3.uiv.p2.T.zoomclick-d3.N-disagree\endcsname{1400.0}
\expandafter\def\csname ppt@rev.desc.d3.uiv.p2.T.zoomclick-d3.N-disagree\endcsname{$+$1400.0}
\expandafter\def\csname val@rev.desc.d3.uiv.p2.T.zoomclick-d3.boot-resid.ci95.0\endcsname{-0.016}
\expandafter\def\csname cnt@rev.desc.d3.uiv.p2.T.zoomclick-d3.boot-resid.ci95.0\endcsname{0}
\expandafter\def\csname pct@rev.desc.d3.uiv.p2.T.zoomclick-d3.boot-resid.ci95.0\endcsname{$-$1.6}
\expandafter\def\csname ppt@rev.desc.d3.uiv.p2.T.zoomclick-d3.boot-resid.ci95.0\endcsname{$-$1.6}
\expandafter\def\csname val@rev.desc.d3.uiv.p2.T.zoomclick-d3.boot-resid.ci95.1\endcsname{0.040}
\expandafter\def\csname cnt@rev.desc.d3.uiv.p2.T.zoomclick-d3.boot-resid.ci95.1\endcsname{0}
\expandafter\def\csname pct@rev.desc.d3.uiv.p2.T.zoomclick-d3.boot-resid.ci95.1\endcsname{4.0}
\expandafter\def\csname ppt@rev.desc.d3.uiv.p2.T.zoomclick-d3.boot-resid.ci95.1\endcsname{$+$4.0}
\expandafter\def\csname val@rev.desc.d3.uiv.p2.T.zoomclick-d3.boot-resid.n-apps\endcsname{24}
\expandafter\def\csname cnt@rev.desc.d3.uiv.p2.T.zoomclick-d3.boot-resid.n-apps\endcsname{24}
\expandafter\def\csname pct@rev.desc.d3.uiv.p2.T.zoomclick-d3.boot-resid.n-apps\endcsname{2400.0}
\expandafter\def\csname ppt@rev.desc.d3.uiv.p2.T.zoomclick-d3.boot-resid.n-apps\endcsname{$+$2400.0}
\expandafter\def\csname val@rev.desc.d3.uiv.p2.T.zoomclick-d3.boot-resid.n-boot-valid\endcsname{4000}
\expandafter\def\csname cnt@rev.desc.d3.uiv.p2.T.zoomclick-d3.boot-resid.n-boot-valid\endcsname{4{,}000}
\expandafter\def\csname pct@rev.desc.d3.uiv.p2.T.zoomclick-d3.boot-resid.n-boot-valid\endcsname{400000.0}
\expandafter\def\csname ppt@rev.desc.d3.uiv.p2.T.zoomclick-d3.boot-resid.n-boot-valid\endcsname{$+$400000.0}
\expandafter\def\csname val@rev.desc.d3.uiv.p2.T.zoomclick-d3.boot-resid.value\endcsname{0.008}
\expandafter\def\csname cnt@rev.desc.d3.uiv.p2.T.zoomclick-d3.boot-resid.value\endcsname{0}
\expandafter\def\csname pct@rev.desc.d3.uiv.p2.T.zoomclick-d3.boot-resid.value\endcsname{0.8}
\expandafter\def\csname ppt@rev.desc.d3.uiv.p2.T.zoomclick-d3.boot-resid.value\endcsname{$+$0.8}
\expandafter\def\csname val@rev.desc.d3.uiv.p2.T.zoomclick-d3.boot-share.ci95.0\endcsname{0.812}
\expandafter\def\csname cnt@rev.desc.d3.uiv.p2.T.zoomclick-d3.boot-share.ci95.0\endcsname{1}
\expandafter\def\csname pct@rev.desc.d3.uiv.p2.T.zoomclick-d3.boot-share.ci95.0\endcsname{81.2}
\expandafter\def\csname ppt@rev.desc.d3.uiv.p2.T.zoomclick-d3.boot-share.ci95.0\endcsname{$+$81.2}
\expandafter\def\csname val@rev.desc.d3.uiv.p2.T.zoomclick-d3.boot-share.ci95.1\endcsname{0.931}
\expandafter\def\csname cnt@rev.desc.d3.uiv.p2.T.zoomclick-d3.boot-share.ci95.1\endcsname{1}
\expandafter\def\csname pct@rev.desc.d3.uiv.p2.T.zoomclick-d3.boot-share.ci95.1\endcsname{93.1}
\expandafter\def\csname ppt@rev.desc.d3.uiv.p2.T.zoomclick-d3.boot-share.ci95.1\endcsname{$+$93.1}
\expandafter\def\csname val@rev.desc.d3.uiv.p2.T.zoomclick-d3.boot-share.n-apps\endcsname{24}
\expandafter\def\csname cnt@rev.desc.d3.uiv.p2.T.zoomclick-d3.boot-share.n-apps\endcsname{24}
\expandafter\def\csname pct@rev.desc.d3.uiv.p2.T.zoomclick-d3.boot-share.n-apps\endcsname{2400.0}
\expandafter\def\csname ppt@rev.desc.d3.uiv.p2.T.zoomclick-d3.boot-share.n-apps\endcsname{$+$2400.0}
\expandafter\def\csname val@rev.desc.d3.uiv.p2.T.zoomclick-d3.boot-share.n-boot-valid\endcsname{4000}
\expandafter\def\csname cnt@rev.desc.d3.uiv.p2.T.zoomclick-d3.boot-share.n-boot-valid\endcsname{4{,}000}
\expandafter\def\csname pct@rev.desc.d3.uiv.p2.T.zoomclick-d3.boot-share.n-boot-valid\endcsname{400000.0}
\expandafter\def\csname ppt@rev.desc.d3.uiv.p2.T.zoomclick-d3.boot-share.n-boot-valid\endcsname{$+$400000.0}
\expandafter\def\csname val@rev.desc.d3.uiv.p2.T.zoomclick-d3.boot-share.value\endcsname{0.875}
\expandafter\def\csname cnt@rev.desc.d3.uiv.p2.T.zoomclick-d3.boot-share.value\endcsname{1}
\expandafter\def\csname pct@rev.desc.d3.uiv.p2.T.zoomclick-d3.boot-share.value\endcsname{87.5}
\expandafter\def\csname ppt@rev.desc.d3.uiv.p2.T.zoomclick-d3.boot-share.value\endcsname{$+$87.5}
\expandafter\def\csname val@rev.desc.d3.uiv.p2.T.zoomclick-d3.n\endcsname{144}
\expandafter\def\csname cnt@rev.desc.d3.uiv.p2.T.zoomclick-d3.n\endcsname{144}
\expandafter\def\csname pct@rev.desc.d3.uiv.p2.T.zoomclick-d3.n\endcsname{14400.0}
\expandafter\def\csname ppt@rev.desc.d3.uiv.p2.T.zoomclick-d3.n\endcsname{$+$14400.0}
\expandafter\def\csname val@rev.desc.d3.uiv.p2.T.zoomclick-d3.reproduces-frozen\endcsname{1}
\expandafter\def\csname cnt@rev.desc.d3.uiv.p2.T.zoomclick-d3.reproduces-frozen\endcsname{1}
\expandafter\def\csname pct@rev.desc.d3.uiv.p2.T.zoomclick-d3.reproduces-frozen\endcsname{100.0}
\expandafter\def\csname ppt@rev.desc.d3.uiv.p2.T.zoomclick-d3.reproduces-frozen\endcsname{$+$100.0}
\expandafter\def\csname val@rev.desc.d3.uiv.p2.T.zoomclick-d3.resid-agree-stable\endcsname{0.008}
\expandafter\def\csname cnt@rev.desc.d3.uiv.p2.T.zoomclick-d3.resid-agree-stable\endcsname{0}
\expandafter\def\csname pct@rev.desc.d3.uiv.p2.T.zoomclick-d3.resid-agree-stable\endcsname{0.8}
\expandafter\def\csname ppt@rev.desc.d3.uiv.p2.T.zoomclick-d3.resid-agree-stable\endcsname{$+$0.8}
\expandafter\def\csname val@rev.desc.d3.uiv.p2.T.zoomclick-d3.resid-disagree\endcsname{0.071}
\expandafter\def\csname cnt@rev.desc.d3.uiv.p2.T.zoomclick-d3.resid-disagree\endcsname{0}
\expandafter\def\csname pct@rev.desc.d3.uiv.p2.T.zoomclick-d3.resid-disagree\endcsname{7.1}
\expandafter\def\csname ppt@rev.desc.d3.uiv.p2.T.zoomclick-d3.resid-disagree\endcsname{$+$7.1}
\expandafter\def\csname val@rev.desc.d3.uiv.p2.T.zoomclick-d3.share-agree-stable\endcsname{0.875}
\expandafter\def\csname cnt@rev.desc.d3.uiv.p2.T.zoomclick-d3.share-agree-stable\endcsname{1}
\expandafter\def\csname pct@rev.desc.d3.uiv.p2.T.zoomclick-d3.share-agree-stable\endcsname{87.5}
\expandafter\def\csname ppt@rev.desc.d3.uiv.p2.T.zoomclick-d3.share-agree-stable\endcsname{$+$87.5}
\expandafter\def\csname val@rev.desc.d4.sspro.A.agree-threshold-over-target-diag.median\endcsname{1.059}
\expandafter\def\csname cnt@rev.desc.d4.sspro.A.agree-threshold-over-target-diag.median\endcsname{1}
\expandafter\def\csname pct@rev.desc.d4.sspro.A.agree-threshold-over-target-diag.median\endcsname{105.9}
\expandafter\def\csname ppt@rev.desc.d4.sspro.A.agree-threshold-over-target-diag.median\endcsname{$+$105.9}
\expandafter\def\csname val@rev.desc.d4.sspro.A.agree-threshold-over-target-diag.n\endcsname{32}
\expandafter\def\csname cnt@rev.desc.d4.sspro.A.agree-threshold-over-target-diag.n\endcsname{32}
\expandafter\def\csname pct@rev.desc.d4.sspro.A.agree-threshold-over-target-diag.n\endcsname{3200.0}
\expandafter\def\csname ppt@rev.desc.d4.sspro.A.agree-threshold-over-target-diag.n\endcsname{$+$3200.0}
\expandafter\def\csname val@rev.desc.d4.sspro.A.agree-threshold-over-target-diag.q25\endcsname{0.495}
\expandafter\def\csname cnt@rev.desc.d4.sspro.A.agree-threshold-over-target-diag.q25\endcsname{0}
\expandafter\def\csname pct@rev.desc.d4.sspro.A.agree-threshold-over-target-diag.q25\endcsname{49.5}
\expandafter\def\csname ppt@rev.desc.d4.sspro.A.agree-threshold-over-target-diag.q25\endcsname{$+$49.5}
\expandafter\def\csname val@rev.desc.d4.sspro.A.agree-threshold-over-target-diag.q75\endcsname{2.075}
\expandafter\def\csname cnt@rev.desc.d4.sspro.A.agree-threshold-over-target-diag.q75\endcsname{2}
\expandafter\def\csname pct@rev.desc.d4.sspro.A.agree-threshold-over-target-diag.q75\endcsname{207.5}
\expandafter\def\csname ppt@rev.desc.d4.sspro.A.agree-threshold-over-target-diag.q75\endcsname{$+$207.5}
\expandafter\def\csname val@rev.desc.d4.sspro.A.agree-threshold-over-target-short-side.median\endcsname{2.996}
\expandafter\def\csname cnt@rev.desc.d4.sspro.A.agree-threshold-over-target-short-side.median\endcsname{3}
\expandafter\def\csname pct@rev.desc.d4.sspro.A.agree-threshold-over-target-short-side.median\endcsname{299.6}
\expandafter\def\csname ppt@rev.desc.d4.sspro.A.agree-threshold-over-target-short-side.median\endcsname{$+$299.6}
\expandafter\def\csname val@rev.desc.d4.sspro.A.agree-threshold-over-target-short-side.n\endcsname{32}
\expandafter\def\csname cnt@rev.desc.d4.sspro.A.agree-threshold-over-target-short-side.n\endcsname{32}
\expandafter\def\csname pct@rev.desc.d4.sspro.A.agree-threshold-over-target-short-side.n\endcsname{3200.0}
\expandafter\def\csname ppt@rev.desc.d4.sspro.A.agree-threshold-over-target-short-side.n\endcsname{$+$3200.0}
\expandafter\def\csname val@rev.desc.d4.sspro.A.agree-threshold-over-target-short-side.q25\endcsname{2.546}
\expandafter\def\csname cnt@rev.desc.d4.sspro.A.agree-threshold-over-target-short-side.q25\endcsname{3}
\expandafter\def\csname pct@rev.desc.d4.sspro.A.agree-threshold-over-target-short-side.q25\endcsname{254.6}
\expandafter\def\csname ppt@rev.desc.d4.sspro.A.agree-threshold-over-target-short-side.q25\endcsname{$+$254.6}
\expandafter\def\csname val@rev.desc.d4.sspro.A.agree-threshold-over-target-short-side.q75\endcsname{4.074}
\expandafter\def\csname cnt@rev.desc.d4.sspro.A.agree-threshold-over-target-short-side.q75\endcsname{4}
\expandafter\def\csname pct@rev.desc.d4.sspro.A.agree-threshold-over-target-short-side.q75\endcsname{407.4}
\expandafter\def\csname ppt@rev.desc.d4.sspro.A.agree-threshold-over-target-short-side.q75\endcsname{$+$407.4}
\expandafter\def\csname val@rev.desc.d4.sspro.A.direct-view-long-edge.median\endcsname{1024.000}
\expandafter\def\csname cnt@rev.desc.d4.sspro.A.direct-view-long-edge.median\endcsname{1{,}024}
\expandafter\def\csname pct@rev.desc.d4.sspro.A.direct-view-long-edge.median\endcsname{102400.0}
\expandafter\def\csname ppt@rev.desc.d4.sspro.A.direct-view-long-edge.median\endcsname{$+$102400.0}
\expandafter\def\csname val@rev.desc.d4.sspro.A.direct-view-long-edge.n\endcsname{32}
\expandafter\def\csname cnt@rev.desc.d4.sspro.A.direct-view-long-edge.n\endcsname{32}
\expandafter\def\csname pct@rev.desc.d4.sspro.A.direct-view-long-edge.n\endcsname{3200.0}
\expandafter\def\csname ppt@rev.desc.d4.sspro.A.direct-view-long-edge.n\endcsname{$+$3200.0}
\expandafter\def\csname val@rev.desc.d4.sspro.A.direct-view-long-edge.q25\endcsname{1024.000}
\expandafter\def\csname cnt@rev.desc.d4.sspro.A.direct-view-long-edge.q25\endcsname{1{,}024}
\expandafter\def\csname pct@rev.desc.d4.sspro.A.direct-view-long-edge.q25\endcsname{102400.0}
\expandafter\def\csname ppt@rev.desc.d4.sspro.A.direct-view-long-edge.q25\endcsname{$+$102400.0}
\expandafter\def\csname val@rev.desc.d4.sspro.A.direct-view-long-edge.q75\endcsname{1024.000}
\expandafter\def\csname cnt@rev.desc.d4.sspro.A.direct-view-long-edge.q75\endcsname{1{,}024}
\expandafter\def\csname pct@rev.desc.d4.sspro.A.direct-view-long-edge.q75\endcsname{102400.0}
\expandafter\def\csname ppt@rev.desc.d4.sspro.A.direct-view-long-edge.q75\endcsname{$+$102400.0}
\expandafter\def\csname val@rev.desc.d4.sspro.A.downscale-factor.median\endcsname{0.400}
\expandafter\def\csname cnt@rev.desc.d4.sspro.A.downscale-factor.median\endcsname{0}
\expandafter\def\csname pct@rev.desc.d4.sspro.A.downscale-factor.median\endcsname{40.0}
\expandafter\def\csname ppt@rev.desc.d4.sspro.A.downscale-factor.median\endcsname{$+$40.0}
\expandafter\def\csname val@rev.desc.d4.sspro.A.downscale-factor.n\endcsname{32}
\expandafter\def\csname cnt@rev.desc.d4.sspro.A.downscale-factor.n\endcsname{32}
\expandafter\def\csname pct@rev.desc.d4.sspro.A.downscale-factor.n\endcsname{3200.0}
\expandafter\def\csname ppt@rev.desc.d4.sspro.A.downscale-factor.n\endcsname{$+$3200.0}
\expandafter\def\csname val@rev.desc.d4.sspro.A.downscale-factor.q25\endcsname{0.296}
\expandafter\def\csname cnt@rev.desc.d4.sspro.A.downscale-factor.q25\endcsname{0}
\expandafter\def\csname pct@rev.desc.d4.sspro.A.downscale-factor.q25\endcsname{29.6}
\expandafter\def\csname ppt@rev.desc.d4.sspro.A.downscale-factor.q25\endcsname{$+$29.6}
\expandafter\def\csname val@rev.desc.d4.sspro.A.downscale-factor.q75\endcsname{0.419}
\expandafter\def\csname cnt@rev.desc.d4.sspro.A.downscale-factor.q75\endcsname{0}
\expandafter\def\csname pct@rev.desc.d4.sspro.A.downscale-factor.q75\endcsname{41.9}
\expandafter\def\csname ppt@rev.desc.d4.sspro.A.downscale-factor.q75\endcsname{$+$41.9}
\expandafter\def\csname val@rev.desc.d4.sspro.A.long-edge.median\endcsname{2560.000}
\expandafter\def\csname cnt@rev.desc.d4.sspro.A.long-edge.median\endcsname{2{,}560}
\expandafter\def\csname pct@rev.desc.d4.sspro.A.long-edge.median\endcsname{256000.0}
\expandafter\def\csname ppt@rev.desc.d4.sspro.A.long-edge.median\endcsname{$+$256000.0}
\expandafter\def\csname val@rev.desc.d4.sspro.A.long-edge.n\endcsname{32}
\expandafter\def\csname cnt@rev.desc.d4.sspro.A.long-edge.n\endcsname{32}
\expandafter\def\csname pct@rev.desc.d4.sspro.A.long-edge.n\endcsname{3200.0}
\expandafter\def\csname ppt@rev.desc.d4.sspro.A.long-edge.n\endcsname{$+$3200.0}
\expandafter\def\csname val@rev.desc.d4.sspro.A.long-edge.q25\endcsname{2460.000}
\expandafter\def\csname cnt@rev.desc.d4.sspro.A.long-edge.q25\endcsname{2{,}460}
\expandafter\def\csname pct@rev.desc.d4.sspro.A.long-edge.q25\endcsname{246000.0}
\expandafter\def\csname ppt@rev.desc.d4.sspro.A.long-edge.q25\endcsname{$+$246000.0}
\expandafter\def\csname val@rev.desc.d4.sspro.A.long-edge.q75\endcsname{3456.000}
\expandafter\def\csname cnt@rev.desc.d4.sspro.A.long-edge.q75\endcsname{3{,}456}
\expandafter\def\csname pct@rev.desc.d4.sspro.A.long-edge.q75\endcsname{345600.0}
\expandafter\def\csname ppt@rev.desc.d4.sspro.A.long-edge.q75\endcsname{$+$345600.0}
\expandafter\def\csname val@rev.desc.d4.sspro.A.megapixels.median\endcsname{3.891}
\expandafter\def\csname cnt@rev.desc.d4.sspro.A.megapixels.median\endcsname{4}
\expandafter\def\csname pct@rev.desc.d4.sspro.A.megapixels.median\endcsname{389.1}
\expandafter\def\csname ppt@rev.desc.d4.sspro.A.megapixels.median\endcsname{$+$389.1}
\expandafter\def\csname val@rev.desc.d4.sspro.A.megapixels.n\endcsname{32}
\expandafter\def\csname cnt@rev.desc.d4.sspro.A.megapixels.n\endcsname{32}
\expandafter\def\csname pct@rev.desc.d4.sspro.A.megapixels.n\endcsname{3200.0}
\expandafter\def\csname ppt@rev.desc.d4.sspro.A.megapixels.n\endcsname{$+$3200.0}
\expandafter\def\csname val@rev.desc.d4.sspro.A.megapixels.q25\endcsname{3.542}
\expandafter\def\csname cnt@rev.desc.d4.sspro.A.megapixels.q25\endcsname{4}
\expandafter\def\csname pct@rev.desc.d4.sspro.A.megapixels.q25\endcsname{354.2}
\expandafter\def\csname ppt@rev.desc.d4.sspro.A.megapixels.q25\endcsname{$+$354.2}
\expandafter\def\csname val@rev.desc.d4.sspro.A.megapixels.q75\endcsname{7.465}
\expandafter\def\csname cnt@rev.desc.d4.sspro.A.megapixels.q75\endcsname{7}
\expandafter\def\csname pct@rev.desc.d4.sspro.A.megapixels.q75\endcsname{746.5}
\expandafter\def\csname ppt@rev.desc.d4.sspro.A.megapixels.q75\endcsname{$+$746.5}
\expandafter\def\csname val@rev.desc.d4.sspro.A.n-apps\endcsname{4}
\expandafter\def\csname cnt@rev.desc.d4.sspro.A.n-apps\endcsname{4}
\expandafter\def\csname pct@rev.desc.d4.sspro.A.n-apps\endcsname{400.0}
\expandafter\def\csname ppt@rev.desc.d4.sspro.A.n-apps\endcsname{$+$400.0}
\expandafter\def\csname val@rev.desc.d4.sspro.A.n-tasks\endcsname{32}
\expandafter\def\csname cnt@rev.desc.d4.sspro.A.n-tasks\endcsname{32}
\expandafter\def\csname pct@rev.desc.d4.sspro.A.n-tasks\endcsname{3200.0}
\expandafter\def\csname ppt@rev.desc.d4.sspro.A.n-tasks\endcsname{$+$3200.0}
\expandafter\def\csname val@rev.desc.d4.sspro.A.share-downscaled\endcsname{1.000}
\expandafter\def\csname cnt@rev.desc.d4.sspro.A.share-downscaled\endcsname{1}
\expandafter\def\csname pct@rev.desc.d4.sspro.A.share-downscaled\endcsname{100.0}
\expandafter\def\csname ppt@rev.desc.d4.sspro.A.share-downscaled\endcsname{$+$100.0}
\expandafter\def\csname val@rev.desc.d4.sspro.A.share-threshold-gt-target-short-side\endcsname{0.938}
\expandafter\def\csname cnt@rev.desc.d4.sspro.A.share-threshold-gt-target-short-side\endcsname{1}
\expandafter\def\csname pct@rev.desc.d4.sspro.A.share-threshold-gt-target-short-side\endcsname{93.8}
\expandafter\def\csname ppt@rev.desc.d4.sspro.A.share-threshold-gt-target-short-side\endcsname{$+$93.8}
\expandafter\def\csname val@rev.desc.d4.sspro.A.short-edge.median\endcsname{1520.000}
\expandafter\def\csname cnt@rev.desc.d4.sspro.A.short-edge.median\endcsname{1{,}520}
\expandafter\def\csname pct@rev.desc.d4.sspro.A.short-edge.median\endcsname{152000.0}
\expandafter\def\csname ppt@rev.desc.d4.sspro.A.short-edge.median\endcsname{$+$152000.0}
\expandafter\def\csname val@rev.desc.d4.sspro.A.short-edge.n\endcsname{32}
\expandafter\def\csname cnt@rev.desc.d4.sspro.A.short-edge.n\endcsname{32}
\expandafter\def\csname pct@rev.desc.d4.sspro.A.short-edge.n\endcsname{3200.0}
\expandafter\def\csname ppt@rev.desc.d4.sspro.A.short-edge.n\endcsname{$+$3200.0}
\expandafter\def\csname val@rev.desc.d4.sspro.A.short-edge.q25\endcsname{1440.000}
\expandafter\def\csname cnt@rev.desc.d4.sspro.A.short-edge.q25\endcsname{1{,}440}
\expandafter\def\csname pct@rev.desc.d4.sspro.A.short-edge.q25\endcsname{144000.0}
\expandafter\def\csname ppt@rev.desc.d4.sspro.A.short-edge.q25\endcsname{$+$144000.0}
\expandafter\def\csname val@rev.desc.d4.sspro.A.short-edge.q75\endcsname{2160.000}
\expandafter\def\csname cnt@rev.desc.d4.sspro.A.short-edge.q75\endcsname{2{,}160}
\expandafter\def\csname pct@rev.desc.d4.sspro.A.short-edge.q75\endcsname{216000.0}
\expandafter\def\csname ppt@rev.desc.d4.sspro.A.short-edge.q75\endcsname{$+$216000.0}
\expandafter\def\csname val@rev.desc.d4.sspro.A.target-area-fraction.median\endcsname{0.000}
\expandafter\def\csname cnt@rev.desc.d4.sspro.A.target-area-fraction.median\endcsname{0}
\expandafter\def\csname pct@rev.desc.d4.sspro.A.target-area-fraction.median\endcsname{0.0}
\expandafter\def\csname ppt@rev.desc.d4.sspro.A.target-area-fraction.median\endcsname{0.0}
\expandafter\def\csname val@rev.desc.d4.sspro.A.target-area-fraction.n\endcsname{32}
\expandafter\def\csname cnt@rev.desc.d4.sspro.A.target-area-fraction.n\endcsname{32}
\expandafter\def\csname pct@rev.desc.d4.sspro.A.target-area-fraction.n\endcsname{3200.0}
\expandafter\def\csname ppt@rev.desc.d4.sspro.A.target-area-fraction.n\endcsname{$+$3200.0}
\expandafter\def\csname val@rev.desc.d4.sspro.A.target-area-fraction.q25\endcsname{0.000}
\expandafter\def\csname cnt@rev.desc.d4.sspro.A.target-area-fraction.q25\endcsname{0}
\expandafter\def\csname pct@rev.desc.d4.sspro.A.target-area-fraction.q25\endcsname{0.0}
\expandafter\def\csname ppt@rev.desc.d4.sspro.A.target-area-fraction.q25\endcsname{0.0}
\expandafter\def\csname val@rev.desc.d4.sspro.A.target-area-fraction.q75\endcsname{0.001}
\expandafter\def\csname cnt@rev.desc.d4.sspro.A.target-area-fraction.q75\endcsname{0}
\expandafter\def\csname pct@rev.desc.d4.sspro.A.target-area-fraction.q75\endcsname{0.1}
\expandafter\def\csname ppt@rev.desc.d4.sspro.A.target-area-fraction.q75\endcsname{$+$0.1}
\expandafter\def\csname val@rev.desc.d4.sspro.A.target-short-side-direct-view.median\endcsname{8.030}
\expandafter\def\csname cnt@rev.desc.d4.sspro.A.target-short-side-direct-view.median\endcsname{8}
\expandafter\def\csname pct@rev.desc.d4.sspro.A.target-short-side-direct-view.median\endcsname{803.0}
\expandafter\def\csname ppt@rev.desc.d4.sspro.A.target-short-side-direct-view.median\endcsname{$+$803.0}
\expandafter\def\csname val@rev.desc.d4.sspro.A.target-short-side-direct-view.n\endcsname{32}
\expandafter\def\csname cnt@rev.desc.d4.sspro.A.target-short-side-direct-view.n\endcsname{32}
\expandafter\def\csname pct@rev.desc.d4.sspro.A.target-short-side-direct-view.n\endcsname{3200.0}
\expandafter\def\csname ppt@rev.desc.d4.sspro.A.target-short-side-direct-view.n\endcsname{$+$3200.0}
\expandafter\def\csname val@rev.desc.d4.sspro.A.target-short-side-direct-view.q25\endcsname{6.030}
\expandafter\def\csname cnt@rev.desc.d4.sspro.A.target-short-side-direct-view.q25\endcsname{6}
\expandafter\def\csname pct@rev.desc.d4.sspro.A.target-short-side-direct-view.q25\endcsname{603.0}
\expandafter\def\csname ppt@rev.desc.d4.sspro.A.target-short-side-direct-view.q25\endcsname{$+$603.0}
\expandafter\def\csname val@rev.desc.d4.sspro.A.target-short-side-direct-view.q75\endcsname{9.296}
\expandafter\def\csname cnt@rev.desc.d4.sspro.A.target-short-side-direct-view.q75\endcsname{9}
\expandafter\def\csname pct@rev.desc.d4.sspro.A.target-short-side-direct-view.q75\endcsname{929.6}
\expandafter\def\csname ppt@rev.desc.d4.sspro.A.target-short-side-direct-view.q75\endcsname{$+$929.6}
\expandafter\def\csname val@rev.desc.d4.sspro.A.target-short-side-native.median\endcsname{22.000}
\expandafter\def\csname cnt@rev.desc.d4.sspro.A.target-short-side-native.median\endcsname{22}
\expandafter\def\csname pct@rev.desc.d4.sspro.A.target-short-side-native.median\endcsname{2200.0}
\expandafter\def\csname ppt@rev.desc.d4.sspro.A.target-short-side-native.median\endcsname{$+$2200.0}
\expandafter\def\csname val@rev.desc.d4.sspro.A.target-short-side-native.n\endcsname{32}
\expandafter\def\csname cnt@rev.desc.d4.sspro.A.target-short-side-native.n\endcsname{32}
\expandafter\def\csname pct@rev.desc.d4.sspro.A.target-short-side-native.n\endcsname{3200.0}
\expandafter\def\csname ppt@rev.desc.d4.sspro.A.target-short-side-native.n\endcsname{$+$3200.0}
\expandafter\def\csname val@rev.desc.d4.sspro.A.target-short-side-native.q25\endcsname{17.000}
\expandafter\def\csname cnt@rev.desc.d4.sspro.A.target-short-side-native.q25\endcsname{17}
\expandafter\def\csname pct@rev.desc.d4.sspro.A.target-short-side-native.q25\endcsname{1700.0}
\expandafter\def\csname ppt@rev.desc.d4.sspro.A.target-short-side-native.q25\endcsname{$+$1700.0}
\expandafter\def\csname val@rev.desc.d4.sspro.A.target-short-side-native.q75\endcsname{29.000}
\expandafter\def\csname cnt@rev.desc.d4.sspro.A.target-short-side-native.q75\endcsname{29}
\expandafter\def\csname pct@rev.desc.d4.sspro.A.target-short-side-native.q75\endcsname{2900.0}
\expandafter\def\csname ppt@rev.desc.d4.sspro.A.target-short-side-native.q75\endcsname{$+$2900.0}
\expandafter\def\csname val@rev.desc.d4.sspro.C.agree-threshold-over-target-diag.median\endcsname{1.303}
\expandafter\def\csname cnt@rev.desc.d4.sspro.C.agree-threshold-over-target-diag.median\endcsname{1}
\expandafter\def\csname pct@rev.desc.d4.sspro.C.agree-threshold-over-target-diag.median\endcsname{130.3}
\expandafter\def\csname ppt@rev.desc.d4.sspro.C.agree-threshold-over-target-diag.median\endcsname{$+$130.3}
\expandafter\def\csname val@rev.desc.d4.sspro.C.agree-threshold-over-target-diag.n\endcsname{48}
\expandafter\def\csname cnt@rev.desc.d4.sspro.C.agree-threshold-over-target-diag.n\endcsname{48}
\expandafter\def\csname pct@rev.desc.d4.sspro.C.agree-threshold-over-target-diag.n\endcsname{4800.0}
\expandafter\def\csname ppt@rev.desc.d4.sspro.C.agree-threshold-over-target-diag.n\endcsname{$+$4800.0}
\expandafter\def\csname val@rev.desc.d4.sspro.C.agree-threshold-over-target-diag.q25\endcsname{0.490}
\expandafter\def\csname cnt@rev.desc.d4.sspro.C.agree-threshold-over-target-diag.q25\endcsname{0}
\expandafter\def\csname pct@rev.desc.d4.sspro.C.agree-threshold-over-target-diag.q25\endcsname{49.0}
\expandafter\def\csname ppt@rev.desc.d4.sspro.C.agree-threshold-over-target-diag.q25\endcsname{$+$49.0}
\expandafter\def\csname val@rev.desc.d4.sspro.C.agree-threshold-over-target-diag.q75\endcsname{2.050}
\expandafter\def\csname cnt@rev.desc.d4.sspro.C.agree-threshold-over-target-diag.q75\endcsname{2}
\expandafter\def\csname pct@rev.desc.d4.sspro.C.agree-threshold-over-target-diag.q75\endcsname{205.0}
\expandafter\def\csname ppt@rev.desc.d4.sspro.C.agree-threshold-over-target-diag.q75\endcsname{$+$205.0}
\expandafter\def\csname val@rev.desc.d4.sspro.C.agree-threshold-over-target-short-side.median\endcsname{3.389}
\expandafter\def\csname cnt@rev.desc.d4.sspro.C.agree-threshold-over-target-short-side.median\endcsname{3}
\expandafter\def\csname pct@rev.desc.d4.sspro.C.agree-threshold-over-target-short-side.median\endcsname{338.9}
\expandafter\def\csname ppt@rev.desc.d4.sspro.C.agree-threshold-over-target-short-side.median\endcsname{$+$338.9}
\expandafter\def\csname val@rev.desc.d4.sspro.C.agree-threshold-over-target-short-side.n\endcsname{48}
\expandafter\def\csname cnt@rev.desc.d4.sspro.C.agree-threshold-over-target-short-side.n\endcsname{48}
\expandafter\def\csname pct@rev.desc.d4.sspro.C.agree-threshold-over-target-short-side.n\endcsname{4800.0}
\expandafter\def\csname ppt@rev.desc.d4.sspro.C.agree-threshold-over-target-short-side.n\endcsname{$+$4800.0}
\expandafter\def\csname val@rev.desc.d4.sspro.C.agree-threshold-over-target-short-side.q25\endcsname{2.238}
\expandafter\def\csname cnt@rev.desc.d4.sspro.C.agree-threshold-over-target-short-side.q25\endcsname{2}
\expandafter\def\csname pct@rev.desc.d4.sspro.C.agree-threshold-over-target-short-side.q25\endcsname{223.8}
\expandafter\def\csname ppt@rev.desc.d4.sspro.C.agree-threshold-over-target-short-side.q25\endcsname{$+$223.8}
\expandafter\def\csname val@rev.desc.d4.sspro.C.agree-threshold-over-target-short-side.q75\endcsname{4.406}
\expandafter\def\csname cnt@rev.desc.d4.sspro.C.agree-threshold-over-target-short-side.q75\endcsname{4}
\expandafter\def\csname pct@rev.desc.d4.sspro.C.agree-threshold-over-target-short-side.q75\endcsname{440.6}
\expandafter\def\csname ppt@rev.desc.d4.sspro.C.agree-threshold-over-target-short-side.q75\endcsname{$+$440.6}
\expandafter\def\csname val@rev.desc.d4.sspro.C.direct-view-long-edge.median\endcsname{1024.000}
\expandafter\def\csname cnt@rev.desc.d4.sspro.C.direct-view-long-edge.median\endcsname{1{,}024}
\expandafter\def\csname pct@rev.desc.d4.sspro.C.direct-view-long-edge.median\endcsname{102400.0}
\expandafter\def\csname ppt@rev.desc.d4.sspro.C.direct-view-long-edge.median\endcsname{$+$102400.0}
\expandafter\def\csname val@rev.desc.d4.sspro.C.direct-view-long-edge.n\endcsname{48}
\expandafter\def\csname cnt@rev.desc.d4.sspro.C.direct-view-long-edge.n\endcsname{48}
\expandafter\def\csname pct@rev.desc.d4.sspro.C.direct-view-long-edge.n\endcsname{4800.0}
\expandafter\def\csname ppt@rev.desc.d4.sspro.C.direct-view-long-edge.n\endcsname{$+$4800.0}
\expandafter\def\csname val@rev.desc.d4.sspro.C.direct-view-long-edge.q25\endcsname{1024.000}
\expandafter\def\csname cnt@rev.desc.d4.sspro.C.direct-view-long-edge.q25\endcsname{1{,}024}
\expandafter\def\csname pct@rev.desc.d4.sspro.C.direct-view-long-edge.q25\endcsname{102400.0}
\expandafter\def\csname ppt@rev.desc.d4.sspro.C.direct-view-long-edge.q25\endcsname{$+$102400.0}
\expandafter\def\csname val@rev.desc.d4.sspro.C.direct-view-long-edge.q75\endcsname{1024.000}
\expandafter\def\csname cnt@rev.desc.d4.sspro.C.direct-view-long-edge.q75\endcsname{1{,}024}
\expandafter\def\csname pct@rev.desc.d4.sspro.C.direct-view-long-edge.q75\endcsname{102400.0}
\expandafter\def\csname ppt@rev.desc.d4.sspro.C.direct-view-long-edge.q75\endcsname{$+$102400.0}
\expandafter\def\csname val@rev.desc.d4.sspro.C.downscale-factor.median\endcsname{0.267}
\expandafter\def\csname cnt@rev.desc.d4.sspro.C.downscale-factor.median\endcsname{0}
\expandafter\def\csname pct@rev.desc.d4.sspro.C.downscale-factor.median\endcsname{26.7}
\expandafter\def\csname ppt@rev.desc.d4.sspro.C.downscale-factor.median\endcsname{$+$26.7}
\expandafter\def\csname val@rev.desc.d4.sspro.C.downscale-factor.n\endcsname{48}
\expandafter\def\csname cnt@rev.desc.d4.sspro.C.downscale-factor.n\endcsname{48}
\expandafter\def\csname pct@rev.desc.d4.sspro.C.downscale-factor.n\endcsname{4800.0}
\expandafter\def\csname ppt@rev.desc.d4.sspro.C.downscale-factor.n\endcsname{$+$4800.0}
\expandafter\def\csname val@rev.desc.d4.sspro.C.downscale-factor.q25\endcsname{0.267}
\expandafter\def\csname cnt@rev.desc.d4.sspro.C.downscale-factor.q25\endcsname{0}
\expandafter\def\csname pct@rev.desc.d4.sspro.C.downscale-factor.q25\endcsname{26.7}
\expandafter\def\csname ppt@rev.desc.d4.sspro.C.downscale-factor.q25\endcsname{$+$26.7}
\expandafter\def\csname val@rev.desc.d4.sspro.C.downscale-factor.q75\endcsname{0.267}
\expandafter\def\csname cnt@rev.desc.d4.sspro.C.downscale-factor.q75\endcsname{0}
\expandafter\def\csname pct@rev.desc.d4.sspro.C.downscale-factor.q75\endcsname{26.7}
\expandafter\def\csname ppt@rev.desc.d4.sspro.C.downscale-factor.q75\endcsname{$+$26.7}
\expandafter\def\csname val@rev.desc.d4.sspro.C.long-edge.median\endcsname{3840.000}
\expandafter\def\csname cnt@rev.desc.d4.sspro.C.long-edge.median\endcsname{3{,}840}
\expandafter\def\csname pct@rev.desc.d4.sspro.C.long-edge.median\endcsname{384000.0}
\expandafter\def\csname ppt@rev.desc.d4.sspro.C.long-edge.median\endcsname{$+$384000.0}
\expandafter\def\csname val@rev.desc.d4.sspro.C.long-edge.n\endcsname{48}
\expandafter\def\csname cnt@rev.desc.d4.sspro.C.long-edge.n\endcsname{48}
\expandafter\def\csname pct@rev.desc.d4.sspro.C.long-edge.n\endcsname{4800.0}
\expandafter\def\csname ppt@rev.desc.d4.sspro.C.long-edge.n\endcsname{$+$4800.0}
\expandafter\def\csname val@rev.desc.d4.sspro.C.long-edge.q25\endcsname{3840.000}
\expandafter\def\csname cnt@rev.desc.d4.sspro.C.long-edge.q25\endcsname{3{,}840}
\expandafter\def\csname pct@rev.desc.d4.sspro.C.long-edge.q25\endcsname{384000.0}
\expandafter\def\csname ppt@rev.desc.d4.sspro.C.long-edge.q25\endcsname{$+$384000.0}
\expandafter\def\csname val@rev.desc.d4.sspro.C.long-edge.q75\endcsname{3840.000}
\expandafter\def\csname cnt@rev.desc.d4.sspro.C.long-edge.q75\endcsname{3{,}840}
\expandafter\def\csname pct@rev.desc.d4.sspro.C.long-edge.q75\endcsname{384000.0}
\expandafter\def\csname ppt@rev.desc.d4.sspro.C.long-edge.q75\endcsname{$+$384000.0}
\expandafter\def\csname val@rev.desc.d4.sspro.C.megapixels.median\endcsname{8.294}
\expandafter\def\csname cnt@rev.desc.d4.sspro.C.megapixels.median\endcsname{8}
\expandafter\def\csname pct@rev.desc.d4.sspro.C.megapixels.median\endcsname{829.4}
\expandafter\def\csname ppt@rev.desc.d4.sspro.C.megapixels.median\endcsname{$+$829.4}
\expandafter\def\csname val@rev.desc.d4.sspro.C.megapixels.n\endcsname{48}
\expandafter\def\csname cnt@rev.desc.d4.sspro.C.megapixels.n\endcsname{48}
\expandafter\def\csname pct@rev.desc.d4.sspro.C.megapixels.n\endcsname{4800.0}
\expandafter\def\csname ppt@rev.desc.d4.sspro.C.megapixels.n\endcsname{$+$4800.0}
\expandafter\def\csname val@rev.desc.d4.sspro.C.megapixels.q25\endcsname{7.721}
\expandafter\def\csname cnt@rev.desc.d4.sspro.C.megapixels.q25\endcsname{8}
\expandafter\def\csname pct@rev.desc.d4.sspro.C.megapixels.q25\endcsname{772.1}
\expandafter\def\csname ppt@rev.desc.d4.sspro.C.megapixels.q25\endcsname{$+$772.1}
\expandafter\def\csname val@rev.desc.d4.sspro.C.megapixels.q75\endcsname{8.294}
\expandafter\def\csname cnt@rev.desc.d4.sspro.C.megapixels.q75\endcsname{8}
\expandafter\def\csname pct@rev.desc.d4.sspro.C.megapixels.q75\endcsname{829.4}
\expandafter\def\csname ppt@rev.desc.d4.sspro.C.megapixels.q75\endcsname{$+$829.4}
\expandafter\def\csname val@rev.desc.d4.sspro.C.n-apps\endcsname{6}
\expandafter\def\csname cnt@rev.desc.d4.sspro.C.n-apps\endcsname{6}
\expandafter\def\csname pct@rev.desc.d4.sspro.C.n-apps\endcsname{600.0}
\expandafter\def\csname ppt@rev.desc.d4.sspro.C.n-apps\endcsname{$+$600.0}
\expandafter\def\csname val@rev.desc.d4.sspro.C.n-tasks\endcsname{48}
\expandafter\def\csname cnt@rev.desc.d4.sspro.C.n-tasks\endcsname{48}
\expandafter\def\csname pct@rev.desc.d4.sspro.C.n-tasks\endcsname{4800.0}
\expandafter\def\csname ppt@rev.desc.d4.sspro.C.n-tasks\endcsname{$+$4800.0}
\expandafter\def\csname val@rev.desc.d4.sspro.C.share-downscaled\endcsname{1.000}
\expandafter\def\csname cnt@rev.desc.d4.sspro.C.share-downscaled\endcsname{1}
\expandafter\def\csname pct@rev.desc.d4.sspro.C.share-downscaled\endcsname{100.0}
\expandafter\def\csname ppt@rev.desc.d4.sspro.C.share-downscaled\endcsname{$+$100.0}
\expandafter\def\csname val@rev.desc.d4.sspro.C.share-threshold-gt-target-short-side\endcsname{0.979}
\expandafter\def\csname cnt@rev.desc.d4.sspro.C.share-threshold-gt-target-short-side\endcsname{1}
\expandafter\def\csname pct@rev.desc.d4.sspro.C.share-threshold-gt-target-short-side\endcsname{97.9}
\expandafter\def\csname ppt@rev.desc.d4.sspro.C.share-threshold-gt-target-short-side\endcsname{$+$97.9}
\expandafter\def\csname val@rev.desc.d4.sspro.C.short-edge.median\endcsname{2160.000}
\expandafter\def\csname cnt@rev.desc.d4.sspro.C.short-edge.median\endcsname{2{,}160}
\expandafter\def\csname pct@rev.desc.d4.sspro.C.short-edge.median\endcsname{216000.0}
\expandafter\def\csname ppt@rev.desc.d4.sspro.C.short-edge.median\endcsname{$+$216000.0}
\expandafter\def\csname val@rev.desc.d4.sspro.C.short-edge.n\endcsname{48}
\expandafter\def\csname cnt@rev.desc.d4.sspro.C.short-edge.n\endcsname{48}
\expandafter\def\csname pct@rev.desc.d4.sspro.C.short-edge.n\endcsname{4800.0}
\expandafter\def\csname ppt@rev.desc.d4.sspro.C.short-edge.n\endcsname{$+$4800.0}
\expandafter\def\csname val@rev.desc.d4.sspro.C.short-edge.q25\endcsname{2160.000}
\expandafter\def\csname cnt@rev.desc.d4.sspro.C.short-edge.q25\endcsname{2{,}160}
\expandafter\def\csname pct@rev.desc.d4.sspro.C.short-edge.q25\endcsname{216000.0}
\expandafter\def\csname ppt@rev.desc.d4.sspro.C.short-edge.q25\endcsname{$+$216000.0}
\expandafter\def\csname val@rev.desc.d4.sspro.C.short-edge.q75\endcsname{2160.000}
\expandafter\def\csname cnt@rev.desc.d4.sspro.C.short-edge.q75\endcsname{2{,}160}
\expandafter\def\csname pct@rev.desc.d4.sspro.C.short-edge.q75\endcsname{216000.0}
\expandafter\def\csname ppt@rev.desc.d4.sspro.C.short-edge.q75\endcsname{$+$216000.0}
\expandafter\def\csname val@rev.desc.d4.sspro.C.target-area-fraction.median\endcsname{0.000}
\expandafter\def\csname cnt@rev.desc.d4.sspro.C.target-area-fraction.median\endcsname{0}
\expandafter\def\csname pct@rev.desc.d4.sspro.C.target-area-fraction.median\endcsname{0.0}
\expandafter\def\csname ppt@rev.desc.d4.sspro.C.target-area-fraction.median\endcsname{0.0}
\expandafter\def\csname val@rev.desc.d4.sspro.C.target-area-fraction.n\endcsname{48}
\expandafter\def\csname cnt@rev.desc.d4.sspro.C.target-area-fraction.n\endcsname{48}
\expandafter\def\csname pct@rev.desc.d4.sspro.C.target-area-fraction.n\endcsname{4800.0}
\expandafter\def\csname ppt@rev.desc.d4.sspro.C.target-area-fraction.n\endcsname{$+$4800.0}
\expandafter\def\csname val@rev.desc.d4.sspro.C.target-area-fraction.q25\endcsname{0.000}
\expandafter\def\csname cnt@rev.desc.d4.sspro.C.target-area-fraction.q25\endcsname{0}
\expandafter\def\csname pct@rev.desc.d4.sspro.C.target-area-fraction.q25\endcsname{0.0}
\expandafter\def\csname ppt@rev.desc.d4.sspro.C.target-area-fraction.q25\endcsname{0.0}
\expandafter\def\csname val@rev.desc.d4.sspro.C.target-area-fraction.q75\endcsname{0.001}
\expandafter\def\csname cnt@rev.desc.d4.sspro.C.target-area-fraction.q75\endcsname{0}
\expandafter\def\csname pct@rev.desc.d4.sspro.C.target-area-fraction.q75\endcsname{0.1}
\expandafter\def\csname ppt@rev.desc.d4.sspro.C.target-area-fraction.q75\endcsname{$+$0.1}
\expandafter\def\csname val@rev.desc.d4.sspro.C.target-short-side-direct-view.median\endcsname{6.933}
\expandafter\def\csname cnt@rev.desc.d4.sspro.C.target-short-side-direct-view.median\endcsname{7}
\expandafter\def\csname pct@rev.desc.d4.sspro.C.target-short-side-direct-view.median\endcsname{693.3}
\expandafter\def\csname ppt@rev.desc.d4.sspro.C.target-short-side-direct-view.median\endcsname{$+$693.3}
\expandafter\def\csname val@rev.desc.d4.sspro.C.target-short-side-direct-view.n\endcsname{48}
\expandafter\def\csname cnt@rev.desc.d4.sspro.C.target-short-side-direct-view.n\endcsname{48}
\expandafter\def\csname pct@rev.desc.d4.sspro.C.target-short-side-direct-view.n\endcsname{4800.0}
\expandafter\def\csname ppt@rev.desc.d4.sspro.C.target-short-side-direct-view.n\endcsname{$+$4800.0}
\expandafter\def\csname val@rev.desc.d4.sspro.C.target-short-side-direct-view.q25\endcsname{5.333}
\expandafter\def\csname cnt@rev.desc.d4.sspro.C.target-short-side-direct-view.q25\endcsname{5}
\expandafter\def\csname pct@rev.desc.d4.sspro.C.target-short-side-direct-view.q25\endcsname{533.3}
\expandafter\def\csname ppt@rev.desc.d4.sspro.C.target-short-side-direct-view.q25\endcsname{$+$533.3}
\expandafter\def\csname val@rev.desc.d4.sspro.C.target-short-side-direct-view.q75\endcsname{10.500}
\expandafter\def\csname cnt@rev.desc.d4.sspro.C.target-short-side-direct-view.q75\endcsname{10}
\expandafter\def\csname pct@rev.desc.d4.sspro.C.target-short-side-direct-view.q75\endcsname{1050.0}
\expandafter\def\csname ppt@rev.desc.d4.sspro.C.target-short-side-direct-view.q75\endcsname{$+$1050.0}
\expandafter\def\csname val@rev.desc.d4.sspro.C.target-short-side-native.median\endcsname{22.500}
\expandafter\def\csname cnt@rev.desc.d4.sspro.C.target-short-side-native.median\endcsname{22}
\expandafter\def\csname pct@rev.desc.d4.sspro.C.target-short-side-native.median\endcsname{2250.0}
\expandafter\def\csname ppt@rev.desc.d4.sspro.C.target-short-side-native.median\endcsname{$+$2250.0}
\expandafter\def\csname val@rev.desc.d4.sspro.C.target-short-side-native.n\endcsname{48}
\expandafter\def\csname cnt@rev.desc.d4.sspro.C.target-short-side-native.n\endcsname{48}
\expandafter\def\csname pct@rev.desc.d4.sspro.C.target-short-side-native.n\endcsname{4800.0}
\expandafter\def\csname ppt@rev.desc.d4.sspro.C.target-short-side-native.n\endcsname{$+$4800.0}
\expandafter\def\csname val@rev.desc.d4.sspro.C.target-short-side-native.q25\endcsname{20.000}
\expandafter\def\csname cnt@rev.desc.d4.sspro.C.target-short-side-native.q25\endcsname{20}
\expandafter\def\csname pct@rev.desc.d4.sspro.C.target-short-side-native.q25\endcsname{2000.0}
\expandafter\def\csname ppt@rev.desc.d4.sspro.C.target-short-side-native.q25\endcsname{$+$2000.0}
\expandafter\def\csname val@rev.desc.d4.sspro.C.target-short-side-native.q75\endcsname{27.250}
\expandafter\def\csname cnt@rev.desc.d4.sspro.C.target-short-side-native.q75\endcsname{27}
\expandafter\def\csname pct@rev.desc.d4.sspro.C.target-short-side-native.q75\endcsname{2725.0}
\expandafter\def\csname ppt@rev.desc.d4.sspro.C.target-short-side-native.q75\endcsname{$+$2725.0}
\expandafter\def\csname val@rev.desc.d4.sspro.T.agree-threshold-over-target-diag.median\endcsname{0.855}
\expandafter\def\csname cnt@rev.desc.d4.sspro.T.agree-threshold-over-target-diag.median\endcsname{1}
\expandafter\def\csname pct@rev.desc.d4.sspro.T.agree-threshold-over-target-diag.median\endcsname{85.5}
\expandafter\def\csname ppt@rev.desc.d4.sspro.T.agree-threshold-over-target-diag.median\endcsname{$+$85.5}
\expandafter\def\csname val@rev.desc.d4.sspro.T.agree-threshold-over-target-diag.n\endcsname{96}
\expandafter\def\csname cnt@rev.desc.d4.sspro.T.agree-threshold-over-target-diag.n\endcsname{96}
\expandafter\def\csname pct@rev.desc.d4.sspro.T.agree-threshold-over-target-diag.n\endcsname{9600.0}
\expandafter\def\csname ppt@rev.desc.d4.sspro.T.agree-threshold-over-target-diag.n\endcsname{$+$9600.0}
\expandafter\def\csname val@rev.desc.d4.sspro.T.agree-threshold-over-target-diag.q25\endcsname{0.360}
\expandafter\def\csname cnt@rev.desc.d4.sspro.T.agree-threshold-over-target-diag.q25\endcsname{0}
\expandafter\def\csname pct@rev.desc.d4.sspro.T.agree-threshold-over-target-diag.q25\endcsname{36.0}
\expandafter\def\csname ppt@rev.desc.d4.sspro.T.agree-threshold-over-target-diag.q25\endcsname{$+$36.0}
\expandafter\def\csname val@rev.desc.d4.sspro.T.agree-threshold-over-target-diag.q75\endcsname{1.694}
\expandafter\def\csname cnt@rev.desc.d4.sspro.T.agree-threshold-over-target-diag.q75\endcsname{2}
\expandafter\def\csname pct@rev.desc.d4.sspro.T.agree-threshold-over-target-diag.q75\endcsname{169.4}
\expandafter\def\csname ppt@rev.desc.d4.sspro.T.agree-threshold-over-target-diag.q75\endcsname{$+$169.4}
\expandafter\def\csname val@rev.desc.d4.sspro.T.agree-threshold-over-target-short-side.median\endcsname{2.743}
\expandafter\def\csname cnt@rev.desc.d4.sspro.T.agree-threshold-over-target-short-side.median\endcsname{3}
\expandafter\def\csname pct@rev.desc.d4.sspro.T.agree-threshold-over-target-short-side.median\endcsname{274.3}
\expandafter\def\csname ppt@rev.desc.d4.sspro.T.agree-threshold-over-target-short-side.median\endcsname{$+$274.3}
\expandafter\def\csname val@rev.desc.d4.sspro.T.agree-threshold-over-target-short-side.n\endcsname{96}
\expandafter\def\csname cnt@rev.desc.d4.sspro.T.agree-threshold-over-target-short-side.n\endcsname{96}
\expandafter\def\csname pct@rev.desc.d4.sspro.T.agree-threshold-over-target-short-side.n\endcsname{9600.0}
\expandafter\def\csname ppt@rev.desc.d4.sspro.T.agree-threshold-over-target-short-side.n\endcsname{$+$9600.0}
\expandafter\def\csname val@rev.desc.d4.sspro.T.agree-threshold-over-target-short-side.q25\endcsname{2.259}
\expandafter\def\csname cnt@rev.desc.d4.sspro.T.agree-threshold-over-target-short-side.q25\endcsname{2}
\expandafter\def\csname pct@rev.desc.d4.sspro.T.agree-threshold-over-target-short-side.q25\endcsname{225.9}
\expandafter\def\csname ppt@rev.desc.d4.sspro.T.agree-threshold-over-target-short-side.q25\endcsname{$+$225.9}
\expandafter\def\csname val@rev.desc.d4.sspro.T.agree-threshold-over-target-short-side.q75\endcsname{3.271}
\expandafter\def\csname cnt@rev.desc.d4.sspro.T.agree-threshold-over-target-short-side.q75\endcsname{3}
\expandafter\def\csname pct@rev.desc.d4.sspro.T.agree-threshold-over-target-short-side.q75\endcsname{327.1}
\expandafter\def\csname ppt@rev.desc.d4.sspro.T.agree-threshold-over-target-short-side.q75\endcsname{$+$327.1}
\expandafter\def\csname val@rev.desc.d4.sspro.T.direct-view-long-edge.median\endcsname{1024.000}
\expandafter\def\csname cnt@rev.desc.d4.sspro.T.direct-view-long-edge.median\endcsname{1{,}024}
\expandafter\def\csname pct@rev.desc.d4.sspro.T.direct-view-long-edge.median\endcsname{102400.0}
\expandafter\def\csname ppt@rev.desc.d4.sspro.T.direct-view-long-edge.median\endcsname{$+$102400.0}
\expandafter\def\csname val@rev.desc.d4.sspro.T.direct-view-long-edge.n\endcsname{96}
\expandafter\def\csname cnt@rev.desc.d4.sspro.T.direct-view-long-edge.n\endcsname{96}
\expandafter\def\csname pct@rev.desc.d4.sspro.T.direct-view-long-edge.n\endcsname{9600.0}
\expandafter\def\csname ppt@rev.desc.d4.sspro.T.direct-view-long-edge.n\endcsname{$+$9600.0}
\expandafter\def\csname val@rev.desc.d4.sspro.T.direct-view-long-edge.q25\endcsname{1024.000}
\expandafter\def\csname cnt@rev.desc.d4.sspro.T.direct-view-long-edge.q25\endcsname{1{,}024}
\expandafter\def\csname pct@rev.desc.d4.sspro.T.direct-view-long-edge.q25\endcsname{102400.0}
\expandafter\def\csname ppt@rev.desc.d4.sspro.T.direct-view-long-edge.q25\endcsname{$+$102400.0}
\expandafter\def\csname val@rev.desc.d4.sspro.T.direct-view-long-edge.q75\endcsname{1024.000}
\expandafter\def\csname cnt@rev.desc.d4.sspro.T.direct-view-long-edge.q75\endcsname{1{,}024}
\expandafter\def\csname pct@rev.desc.d4.sspro.T.direct-view-long-edge.q75\endcsname{102400.0}
\expandafter\def\csname ppt@rev.desc.d4.sspro.T.direct-view-long-edge.q75\endcsname{$+$102400.0}
\expandafter\def\csname val@rev.desc.d4.sspro.T.downscale-factor.median\endcsname{0.356}
\expandafter\def\csname cnt@rev.desc.d4.sspro.T.downscale-factor.median\endcsname{0}
\expandafter\def\csname pct@rev.desc.d4.sspro.T.downscale-factor.median\endcsname{35.6}
\expandafter\def\csname ppt@rev.desc.d4.sspro.T.downscale-factor.median\endcsname{$+$35.6}
\expandafter\def\csname val@rev.desc.d4.sspro.T.downscale-factor.n\endcsname{96}
\expandafter\def\csname cnt@rev.desc.d4.sspro.T.downscale-factor.n\endcsname{96}
\expandafter\def\csname pct@rev.desc.d4.sspro.T.downscale-factor.n\endcsname{9600.0}
\expandafter\def\csname ppt@rev.desc.d4.sspro.T.downscale-factor.n\endcsname{$+$9600.0}
\expandafter\def\csname val@rev.desc.d4.sspro.T.downscale-factor.q25\endcsname{0.289}
\expandafter\def\csname cnt@rev.desc.d4.sspro.T.downscale-factor.q25\endcsname{0}
\expandafter\def\csname pct@rev.desc.d4.sspro.T.downscale-factor.q25\endcsname{28.9}
\expandafter\def\csname ppt@rev.desc.d4.sspro.T.downscale-factor.q25\endcsname{$+$28.9}
\expandafter\def\csname val@rev.desc.d4.sspro.T.downscale-factor.q75\endcsname{0.400}
\expandafter\def\csname cnt@rev.desc.d4.sspro.T.downscale-factor.q75\endcsname{0}
\expandafter\def\csname pct@rev.desc.d4.sspro.T.downscale-factor.q75\endcsname{40.0}
\expandafter\def\csname ppt@rev.desc.d4.sspro.T.downscale-factor.q75\endcsname{$+$40.0}
\expandafter\def\csname val@rev.desc.d4.sspro.T.long-edge.median\endcsname{2880.000}
\expandafter\def\csname cnt@rev.desc.d4.sspro.T.long-edge.median\endcsname{2{,}880}
\expandafter\def\csname pct@rev.desc.d4.sspro.T.long-edge.median\endcsname{288000.0}
\expandafter\def\csname ppt@rev.desc.d4.sspro.T.long-edge.median\endcsname{$+$288000.0}
\expandafter\def\csname val@rev.desc.d4.sspro.T.long-edge.n\endcsname{96}
\expandafter\def\csname cnt@rev.desc.d4.sspro.T.long-edge.n\endcsname{96}
\expandafter\def\csname pct@rev.desc.d4.sspro.T.long-edge.n\endcsname{9600.0}
\expandafter\def\csname ppt@rev.desc.d4.sspro.T.long-edge.n\endcsname{$+$9600.0}
\expandafter\def\csname val@rev.desc.d4.sspro.T.long-edge.q25\endcsname{2560.000}
\expandafter\def\csname cnt@rev.desc.d4.sspro.T.long-edge.q25\endcsname{2{,}560}
\expandafter\def\csname pct@rev.desc.d4.sspro.T.long-edge.q25\endcsname{256000.0}
\expandafter\def\csname ppt@rev.desc.d4.sspro.T.long-edge.q25\endcsname{$+$256000.0}
\expandafter\def\csname val@rev.desc.d4.sspro.T.long-edge.q75\endcsname{3552.000}
\expandafter\def\csname cnt@rev.desc.d4.sspro.T.long-edge.q75\endcsname{3{,}552}
\expandafter\def\csname pct@rev.desc.d4.sspro.T.long-edge.q75\endcsname{355200.0}
\expandafter\def\csname ppt@rev.desc.d4.sspro.T.long-edge.q75\endcsname{$+$355200.0}
\expandafter\def\csname val@rev.desc.d4.sspro.T.megapixels.median\endcsname{4.147}
\expandafter\def\csname cnt@rev.desc.d4.sspro.T.megapixels.median\endcsname{4}
\expandafter\def\csname pct@rev.desc.d4.sspro.T.megapixels.median\endcsname{414.7}
\expandafter\def\csname ppt@rev.desc.d4.sspro.T.megapixels.median\endcsname{$+$414.7}
\expandafter\def\csname val@rev.desc.d4.sspro.T.megapixels.n\endcsname{96}
\expandafter\def\csname cnt@rev.desc.d4.sspro.T.megapixels.n\endcsname{96}
\expandafter\def\csname pct@rev.desc.d4.sspro.T.megapixels.n\endcsname{9600.0}
\expandafter\def\csname ppt@rev.desc.d4.sspro.T.megapixels.n\endcsname{$+$9600.0}
\expandafter\def\csname val@rev.desc.d4.sspro.T.megapixels.q25\endcsname{3.686}
\expandafter\def\csname cnt@rev.desc.d4.sspro.T.megapixels.q25\endcsname{4}
\expandafter\def\csname pct@rev.desc.d4.sspro.T.megapixels.q25\endcsname{368.6}
\expandafter\def\csname ppt@rev.desc.d4.sspro.T.megapixels.q25\endcsname{$+$368.6}
\expandafter\def\csname val@rev.desc.d4.sspro.T.megapixels.q75\endcsname{6.126}
\expandafter\def\csname cnt@rev.desc.d4.sspro.T.megapixels.q75\endcsname{6}
\expandafter\def\csname pct@rev.desc.d4.sspro.T.megapixels.q75\endcsname{612.6}
\expandafter\def\csname ppt@rev.desc.d4.sspro.T.megapixels.q75\endcsname{$+$612.6}
\expandafter\def\csname val@rev.desc.d4.sspro.T.n-apps\endcsname{12}
\expandafter\def\csname cnt@rev.desc.d4.sspro.T.n-apps\endcsname{12}
\expandafter\def\csname pct@rev.desc.d4.sspro.T.n-apps\endcsname{1200.0}
\expandafter\def\csname ppt@rev.desc.d4.sspro.T.n-apps\endcsname{$+$1200.0}
\expandafter\def\csname val@rev.desc.d4.sspro.T.n-tasks\endcsname{96}
\expandafter\def\csname cnt@rev.desc.d4.sspro.T.n-tasks\endcsname{96}
\expandafter\def\csname pct@rev.desc.d4.sspro.T.n-tasks\endcsname{9600.0}
\expandafter\def\csname ppt@rev.desc.d4.sspro.T.n-tasks\endcsname{$+$9600.0}
\expandafter\def\csname val@rev.desc.d4.sspro.T.share-downscaled\endcsname{1.000}
\expandafter\def\csname cnt@rev.desc.d4.sspro.T.share-downscaled\endcsname{1}
\expandafter\def\csname pct@rev.desc.d4.sspro.T.share-downscaled\endcsname{100.0}
\expandafter\def\csname ppt@rev.desc.d4.sspro.T.share-downscaled\endcsname{$+$100.0}
\expandafter\def\csname val@rev.desc.d4.sspro.T.share-threshold-gt-target-short-side\endcsname{1.000}
\expandafter\def\csname cnt@rev.desc.d4.sspro.T.share-threshold-gt-target-short-side\endcsname{1}
\expandafter\def\csname pct@rev.desc.d4.sspro.T.share-threshold-gt-target-short-side\endcsname{100.0}
\expandafter\def\csname ppt@rev.desc.d4.sspro.T.share-threshold-gt-target-short-side\endcsname{$+$100.0}
\expandafter\def\csname val@rev.desc.d4.sspro.T.short-edge.median\endcsname{1440.000}
\expandafter\def\csname cnt@rev.desc.d4.sspro.T.short-edge.median\endcsname{1{,}440}
\expandafter\def\csname pct@rev.desc.d4.sspro.T.short-edge.median\endcsname{144000.0}
\expandafter\def\csname ppt@rev.desc.d4.sspro.T.short-edge.median\endcsname{$+$144000.0}
\expandafter\def\csname val@rev.desc.d4.sspro.T.short-edge.n\endcsname{96}
\expandafter\def\csname cnt@rev.desc.d4.sspro.T.short-edge.n\endcsname{96}
\expandafter\def\csname pct@rev.desc.d4.sspro.T.short-edge.n\endcsname{9600.0}
\expandafter\def\csname ppt@rev.desc.d4.sspro.T.short-edge.n\endcsname{$+$9600.0}
\expandafter\def\csname val@rev.desc.d4.sspro.T.short-edge.q25\endcsname{1440.000}
\expandafter\def\csname cnt@rev.desc.d4.sspro.T.short-edge.q25\endcsname{1{,}440}
\expandafter\def\csname pct@rev.desc.d4.sspro.T.short-edge.q25\endcsname{144000.0}
\expandafter\def\csname ppt@rev.desc.d4.sspro.T.short-edge.q25\endcsname{$+$144000.0}
\expandafter\def\csname val@rev.desc.d4.sspro.T.short-edge.q75\endcsname{1961.000}
\expandafter\def\csname cnt@rev.desc.d4.sspro.T.short-edge.q75\endcsname{1{,}961}
\expandafter\def\csname pct@rev.desc.d4.sspro.T.short-edge.q75\endcsname{196100.0}
\expandafter\def\csname ppt@rev.desc.d4.sspro.T.short-edge.q75\endcsname{$+$196100.0}
\expandafter\def\csname val@rev.desc.d4.sspro.T.target-area-fraction.median\endcsname{0.000}
\expandafter\def\csname cnt@rev.desc.d4.sspro.T.target-area-fraction.median\endcsname{0}
\expandafter\def\csname pct@rev.desc.d4.sspro.T.target-area-fraction.median\endcsname{0.0}
\expandafter\def\csname ppt@rev.desc.d4.sspro.T.target-area-fraction.median\endcsname{0.0}
\expandafter\def\csname val@rev.desc.d4.sspro.T.target-area-fraction.n\endcsname{96}
\expandafter\def\csname cnt@rev.desc.d4.sspro.T.target-area-fraction.n\endcsname{96}
\expandafter\def\csname pct@rev.desc.d4.sspro.T.target-area-fraction.n\endcsname{9600.0}
\expandafter\def\csname ppt@rev.desc.d4.sspro.T.target-area-fraction.n\endcsname{$+$9600.0}
\expandafter\def\csname val@rev.desc.d4.sspro.T.target-area-fraction.q25\endcsname{0.000}
\expandafter\def\csname cnt@rev.desc.d4.sspro.T.target-area-fraction.q25\endcsname{0}
\expandafter\def\csname pct@rev.desc.d4.sspro.T.target-area-fraction.q25\endcsname{0.0}
\expandafter\def\csname ppt@rev.desc.d4.sspro.T.target-area-fraction.q25\endcsname{0.0}
\expandafter\def\csname val@rev.desc.d4.sspro.T.target-area-fraction.q75\endcsname{0.001}
\expandafter\def\csname cnt@rev.desc.d4.sspro.T.target-area-fraction.q75\endcsname{0}
\expandafter\def\csname pct@rev.desc.d4.sspro.T.target-area-fraction.q75\endcsname{0.1}
\expandafter\def\csname ppt@rev.desc.d4.sspro.T.target-area-fraction.q75\endcsname{$+$0.1}
\expandafter\def\csname val@rev.desc.d4.sspro.T.target-short-side-direct-view.median\endcsname{8.800}
\expandafter\def\csname cnt@rev.desc.d4.sspro.T.target-short-side-direct-view.median\endcsname{9}
\expandafter\def\csname pct@rev.desc.d4.sspro.T.target-short-side-direct-view.median\endcsname{880.0}
\expandafter\def\csname ppt@rev.desc.d4.sspro.T.target-short-side-direct-view.median\endcsname{$+$880.0}
\expandafter\def\csname val@rev.desc.d4.sspro.T.target-short-side-direct-view.n\endcsname{96}
\expandafter\def\csname cnt@rev.desc.d4.sspro.T.target-short-side-direct-view.n\endcsname{96}
\expandafter\def\csname pct@rev.desc.d4.sspro.T.target-short-side-direct-view.n\endcsname{9600.0}
\expandafter\def\csname ppt@rev.desc.d4.sspro.T.target-short-side-direct-view.n\endcsname{$+$9600.0}
\expandafter\def\csname val@rev.desc.d4.sspro.T.target-short-side-direct-view.q25\endcsname{6.811}
\expandafter\def\csname cnt@rev.desc.d4.sspro.T.target-short-side-direct-view.q25\endcsname{7}
\expandafter\def\csname pct@rev.desc.d4.sspro.T.target-short-side-direct-view.q25\endcsname{681.1}
\expandafter\def\csname ppt@rev.desc.d4.sspro.T.target-short-side-direct-view.q25\endcsname{$+$681.1}
\expandafter\def\csname val@rev.desc.d4.sspro.T.target-short-side-direct-view.q75\endcsname{10.400}
\expandafter\def\csname cnt@rev.desc.d4.sspro.T.target-short-side-direct-view.q75\endcsname{10}
\expandafter\def\csname pct@rev.desc.d4.sspro.T.target-short-side-direct-view.q75\endcsname{1040.0}
\expandafter\def\csname ppt@rev.desc.d4.sspro.T.target-short-side-direct-view.q75\endcsname{$+$1040.0}
\expandafter\def\csname val@rev.desc.d4.sspro.T.target-short-side-native.median\endcsname{24.000}
\expandafter\def\csname cnt@rev.desc.d4.sspro.T.target-short-side-native.median\endcsname{24}
\expandafter\def\csname pct@rev.desc.d4.sspro.T.target-short-side-native.median\endcsname{2400.0}
\expandafter\def\csname ppt@rev.desc.d4.sspro.T.target-short-side-native.median\endcsname{$+$2400.0}
\expandafter\def\csname val@rev.desc.d4.sspro.T.target-short-side-native.n\endcsname{96}
\expandafter\def\csname cnt@rev.desc.d4.sspro.T.target-short-side-native.n\endcsname{96}
\expandafter\def\csname pct@rev.desc.d4.sspro.T.target-short-side-native.n\endcsname{9600.0}
\expandafter\def\csname ppt@rev.desc.d4.sspro.T.target-short-side-native.n\endcsname{$+$9600.0}
\expandafter\def\csname val@rev.desc.d4.sspro.T.target-short-side-native.q25\endcsname{19.750}
\expandafter\def\csname cnt@rev.desc.d4.sspro.T.target-short-side-native.q25\endcsname{20}
\expandafter\def\csname pct@rev.desc.d4.sspro.T.target-short-side-native.q25\endcsname{1975.0}
\expandafter\def\csname ppt@rev.desc.d4.sspro.T.target-short-side-native.q25\endcsname{$+$1975.0}
\expandafter\def\csname val@rev.desc.d4.sspro.T.target-short-side-native.q75\endcsname{31.250}
\expandafter\def\csname cnt@rev.desc.d4.sspro.T.target-short-side-native.q75\endcsname{31}
\expandafter\def\csname pct@rev.desc.d4.sspro.T.target-short-side-native.q75\endcsname{3125.0}
\expandafter\def\csname ppt@rev.desc.d4.sspro.T.target-short-side-native.q75\endcsname{$+$3125.0}
\expandafter\def\csname val@rev.desc.d4.sspro.all.agree-threshold-over-target-diag.median\endcsname{0.894}
\expandafter\def\csname cnt@rev.desc.d4.sspro.all.agree-threshold-over-target-diag.median\endcsname{1}
\expandafter\def\csname pct@rev.desc.d4.sspro.all.agree-threshold-over-target-diag.median\endcsname{89.4}
\expandafter\def\csname ppt@rev.desc.d4.sspro.all.agree-threshold-over-target-diag.median\endcsname{$+$89.4}
\expandafter\def\csname val@rev.desc.d4.sspro.all.agree-threshold-over-target-diag.n\endcsname{176}
\expandafter\def\csname cnt@rev.desc.d4.sspro.all.agree-threshold-over-target-diag.n\endcsname{176}
\expandafter\def\csname pct@rev.desc.d4.sspro.all.agree-threshold-over-target-diag.n\endcsname{17600.0}
\expandafter\def\csname ppt@rev.desc.d4.sspro.all.agree-threshold-over-target-diag.n\endcsname{$+$17600.0}
\expandafter\def\csname val@rev.desc.d4.sspro.all.agree-threshold-over-target-diag.q25\endcsname{0.399}
\expandafter\def\csname cnt@rev.desc.d4.sspro.all.agree-threshold-over-target-diag.q25\endcsname{0}
\expandafter\def\csname pct@rev.desc.d4.sspro.all.agree-threshold-over-target-diag.q25\endcsname{39.9}
\expandafter\def\csname ppt@rev.desc.d4.sspro.all.agree-threshold-over-target-diag.q25\endcsname{$+$39.9}
\expandafter\def\csname val@rev.desc.d4.sspro.all.agree-threshold-over-target-diag.q75\endcsname{1.856}
\expandafter\def\csname cnt@rev.desc.d4.sspro.all.agree-threshold-over-target-diag.q75\endcsname{2}
\expandafter\def\csname pct@rev.desc.d4.sspro.all.agree-threshold-over-target-diag.q75\endcsname{185.6}
\expandafter\def\csname ppt@rev.desc.d4.sspro.all.agree-threshold-over-target-diag.q75\endcsname{$+$185.6}
\expandafter\def\csname val@rev.desc.d4.sspro.all.agree-threshold-over-target-short-side.median\endcsname{2.911}
\expandafter\def\csname cnt@rev.desc.d4.sspro.all.agree-threshold-over-target-short-side.median\endcsname{3}
\expandafter\def\csname pct@rev.desc.d4.sspro.all.agree-threshold-over-target-short-side.median\endcsname{291.1}
\expandafter\def\csname ppt@rev.desc.d4.sspro.all.agree-threshold-over-target-short-side.median\endcsname{$+$291.1}
\expandafter\def\csname val@rev.desc.d4.sspro.all.agree-threshold-over-target-short-side.n\endcsname{176}
\expandafter\def\csname cnt@rev.desc.d4.sspro.all.agree-threshold-over-target-short-side.n\endcsname{176}
\expandafter\def\csname pct@rev.desc.d4.sspro.all.agree-threshold-over-target-short-side.n\endcsname{17600.0}
\expandafter\def\csname ppt@rev.desc.d4.sspro.all.agree-threshold-over-target-short-side.n\endcsname{$+$17600.0}
\expandafter\def\csname val@rev.desc.d4.sspro.all.agree-threshold-over-target-short-side.q25\endcsname{2.259}
\expandafter\def\csname cnt@rev.desc.d4.sspro.all.agree-threshold-over-target-short-side.q25\endcsname{2}
\expandafter\def\csname pct@rev.desc.d4.sspro.all.agree-threshold-over-target-short-side.q25\endcsname{225.9}
\expandafter\def\csname ppt@rev.desc.d4.sspro.all.agree-threshold-over-target-short-side.q25\endcsname{$+$225.9}
\expandafter\def\csname val@rev.desc.d4.sspro.all.agree-threshold-over-target-short-side.q75\endcsname{3.612}
\expandafter\def\csname cnt@rev.desc.d4.sspro.all.agree-threshold-over-target-short-side.q75\endcsname{4}
\expandafter\def\csname pct@rev.desc.d4.sspro.all.agree-threshold-over-target-short-side.q75\endcsname{361.2}
\expandafter\def\csname ppt@rev.desc.d4.sspro.all.agree-threshold-over-target-short-side.q75\endcsname{$+$361.2}
\expandafter\def\csname val@rev.desc.d4.sspro.all.direct-view-long-edge.median\endcsname{1024.000}
\expandafter\def\csname cnt@rev.desc.d4.sspro.all.direct-view-long-edge.median\endcsname{1{,}024}
\expandafter\def\csname pct@rev.desc.d4.sspro.all.direct-view-long-edge.median\endcsname{102400.0}
\expandafter\def\csname ppt@rev.desc.d4.sspro.all.direct-view-long-edge.median\endcsname{$+$102400.0}
\expandafter\def\csname val@rev.desc.d4.sspro.all.direct-view-long-edge.n\endcsname{176}
\expandafter\def\csname cnt@rev.desc.d4.sspro.all.direct-view-long-edge.n\endcsname{176}
\expandafter\def\csname pct@rev.desc.d4.sspro.all.direct-view-long-edge.n\endcsname{17600.0}
\expandafter\def\csname ppt@rev.desc.d4.sspro.all.direct-view-long-edge.n\endcsname{$+$17600.0}
\expandafter\def\csname val@rev.desc.d4.sspro.all.direct-view-long-edge.q25\endcsname{1024.000}
\expandafter\def\csname cnt@rev.desc.d4.sspro.all.direct-view-long-edge.q25\endcsname{1{,}024}
\expandafter\def\csname pct@rev.desc.d4.sspro.all.direct-view-long-edge.q25\endcsname{102400.0}
\expandafter\def\csname ppt@rev.desc.d4.sspro.all.direct-view-long-edge.q25\endcsname{$+$102400.0}
\expandafter\def\csname val@rev.desc.d4.sspro.all.direct-view-long-edge.q75\endcsname{1024.000}
\expandafter\def\csname cnt@rev.desc.d4.sspro.all.direct-view-long-edge.q75\endcsname{1{,}024}
\expandafter\def\csname pct@rev.desc.d4.sspro.all.direct-view-long-edge.q75\endcsname{102400.0}
\expandafter\def\csname ppt@rev.desc.d4.sspro.all.direct-view-long-edge.q75\endcsname{$+$102400.0}
\expandafter\def\csname val@rev.desc.d4.sspro.all.downscale-factor.median\endcsname{0.296}
\expandafter\def\csname cnt@rev.desc.d4.sspro.all.downscale-factor.median\endcsname{0}
\expandafter\def\csname pct@rev.desc.d4.sspro.all.downscale-factor.median\endcsname{29.6}
\expandafter\def\csname ppt@rev.desc.d4.sspro.all.downscale-factor.median\endcsname{$+$29.6}
\expandafter\def\csname val@rev.desc.d4.sspro.all.downscale-factor.n\endcsname{176}
\expandafter\def\csname cnt@rev.desc.d4.sspro.all.downscale-factor.n\endcsname{176}
\expandafter\def\csname pct@rev.desc.d4.sspro.all.downscale-factor.n\endcsname{17600.0}
\expandafter\def\csname ppt@rev.desc.d4.sspro.all.downscale-factor.n\endcsname{$+$17600.0}
\expandafter\def\csname val@rev.desc.d4.sspro.all.downscale-factor.q25\endcsname{0.267}
\expandafter\def\csname cnt@rev.desc.d4.sspro.all.downscale-factor.q25\endcsname{0}
\expandafter\def\csname pct@rev.desc.d4.sspro.all.downscale-factor.q25\endcsname{26.7}
\expandafter\def\csname ppt@rev.desc.d4.sspro.all.downscale-factor.q25\endcsname{$+$26.7}
\expandafter\def\csname val@rev.desc.d4.sspro.all.downscale-factor.q75\endcsname{0.400}
\expandafter\def\csname cnt@rev.desc.d4.sspro.all.downscale-factor.q75\endcsname{0}
\expandafter\def\csname pct@rev.desc.d4.sspro.all.downscale-factor.q75\endcsname{40.0}
\expandafter\def\csname ppt@rev.desc.d4.sspro.all.downscale-factor.q75\endcsname{$+$40.0}
\expandafter\def\csname val@rev.desc.d4.sspro.all.long-edge.median\endcsname{3456.000}
\expandafter\def\csname cnt@rev.desc.d4.sspro.all.long-edge.median\endcsname{3{,}456}
\expandafter\def\csname pct@rev.desc.d4.sspro.all.long-edge.median\endcsname{345600.0}
\expandafter\def\csname ppt@rev.desc.d4.sspro.all.long-edge.median\endcsname{$+$345600.0}
\expandafter\def\csname val@rev.desc.d4.sspro.all.long-edge.n\endcsname{176}
\expandafter\def\csname cnt@rev.desc.d4.sspro.all.long-edge.n\endcsname{176}
\expandafter\def\csname pct@rev.desc.d4.sspro.all.long-edge.n\endcsname{17600.0}
\expandafter\def\csname ppt@rev.desc.d4.sspro.all.long-edge.n\endcsname{$+$17600.0}
\expandafter\def\csname val@rev.desc.d4.sspro.all.long-edge.q25\endcsname{2560.000}
\expandafter\def\csname cnt@rev.desc.d4.sspro.all.long-edge.q25\endcsname{2{,}560}
\expandafter\def\csname pct@rev.desc.d4.sspro.all.long-edge.q25\endcsname{256000.0}
\expandafter\def\csname ppt@rev.desc.d4.sspro.all.long-edge.q25\endcsname{$+$256000.0}
\expandafter\def\csname val@rev.desc.d4.sspro.all.long-edge.q75\endcsname{3840.000}
\expandafter\def\csname cnt@rev.desc.d4.sspro.all.long-edge.q75\endcsname{3{,}840}
\expandafter\def\csname pct@rev.desc.d4.sspro.all.long-edge.q75\endcsname{384000.0}
\expandafter\def\csname ppt@rev.desc.d4.sspro.all.long-edge.q75\endcsname{$+$384000.0}
\expandafter\def\csname val@rev.desc.d4.sspro.all.megapixels.median\endcsname{4.722}
\expandafter\def\csname cnt@rev.desc.d4.sspro.all.megapixels.median\endcsname{5}
\expandafter\def\csname pct@rev.desc.d4.sspro.all.megapixels.median\endcsname{472.2}
\expandafter\def\csname ppt@rev.desc.d4.sspro.all.megapixels.median\endcsname{$+$472.2}
\expandafter\def\csname val@rev.desc.d4.sspro.all.megapixels.n\endcsname{176}
\expandafter\def\csname cnt@rev.desc.d4.sspro.all.megapixels.n\endcsname{176}
\expandafter\def\csname pct@rev.desc.d4.sspro.all.megapixels.n\endcsname{17600.0}
\expandafter\def\csname ppt@rev.desc.d4.sspro.all.megapixels.n\endcsname{$+$17600.0}
\expandafter\def\csname val@rev.desc.d4.sspro.all.megapixels.q25\endcsname{3.686}
\expandafter\def\csname cnt@rev.desc.d4.sspro.all.megapixels.q25\endcsname{4}
\expandafter\def\csname pct@rev.desc.d4.sspro.all.megapixels.q25\endcsname{368.6}
\expandafter\def\csname ppt@rev.desc.d4.sspro.all.megapixels.q25\endcsname{$+$368.6}
\expandafter\def\csname val@rev.desc.d4.sspro.all.megapixels.q75\endcsname{7.864}
\expandafter\def\csname cnt@rev.desc.d4.sspro.all.megapixels.q75\endcsname{8}
\expandafter\def\csname pct@rev.desc.d4.sspro.all.megapixels.q75\endcsname{786.4}
\expandafter\def\csname ppt@rev.desc.d4.sspro.all.megapixels.q75\endcsname{$+$786.4}
\expandafter\def\csname val@rev.desc.d4.sspro.all.n-apps\endcsname{22}
\expandafter\def\csname cnt@rev.desc.d4.sspro.all.n-apps\endcsname{22}
\expandafter\def\csname pct@rev.desc.d4.sspro.all.n-apps\endcsname{2200.0}
\expandafter\def\csname ppt@rev.desc.d4.sspro.all.n-apps\endcsname{$+$2200.0}
\expandafter\def\csname val@rev.desc.d4.sspro.all.n-tasks\endcsname{176}
\expandafter\def\csname cnt@rev.desc.d4.sspro.all.n-tasks\endcsname{176}
\expandafter\def\csname pct@rev.desc.d4.sspro.all.n-tasks\endcsname{17600.0}
\expandafter\def\csname ppt@rev.desc.d4.sspro.all.n-tasks\endcsname{$+$17600.0}
\expandafter\def\csname val@rev.desc.d4.sspro.all.share-downscaled\endcsname{1.000}
\expandafter\def\csname cnt@rev.desc.d4.sspro.all.share-downscaled\endcsname{1}
\expandafter\def\csname pct@rev.desc.d4.sspro.all.share-downscaled\endcsname{100.0}
\expandafter\def\csname ppt@rev.desc.d4.sspro.all.share-downscaled\endcsname{$+$100.0}
\expandafter\def\csname val@rev.desc.d4.sspro.all.share-threshold-gt-target-short-side\endcsname{0.983}
\expandafter\def\csname cnt@rev.desc.d4.sspro.all.share-threshold-gt-target-short-side\endcsname{1}
\expandafter\def\csname pct@rev.desc.d4.sspro.all.share-threshold-gt-target-short-side\endcsname{98.3}
\expandafter\def\csname ppt@rev.desc.d4.sspro.all.share-threshold-gt-target-short-side\endcsname{$+$98.3}
\expandafter\def\csname val@rev.desc.d4.sspro.all.short-edge.median\endcsname{1664.000}
\expandafter\def\csname cnt@rev.desc.d4.sspro.all.short-edge.median\endcsname{1{,}664}
\expandafter\def\csname pct@rev.desc.d4.sspro.all.short-edge.median\endcsname{166400.0}
\expandafter\def\csname ppt@rev.desc.d4.sspro.all.short-edge.median\endcsname{$+$166400.0}
\expandafter\def\csname val@rev.desc.d4.sspro.all.short-edge.n\endcsname{176}
\expandafter\def\csname cnt@rev.desc.d4.sspro.all.short-edge.n\endcsname{176}
\expandafter\def\csname pct@rev.desc.d4.sspro.all.short-edge.n\endcsname{17600.0}
\expandafter\def\csname ppt@rev.desc.d4.sspro.all.short-edge.n\endcsname{$+$17600.0}
\expandafter\def\csname val@rev.desc.d4.sspro.all.short-edge.q25\endcsname{1440.000}
\expandafter\def\csname cnt@rev.desc.d4.sspro.all.short-edge.q25\endcsname{1{,}440}
\expandafter\def\csname pct@rev.desc.d4.sspro.all.short-edge.q25\endcsname{144000.0}
\expandafter\def\csname ppt@rev.desc.d4.sspro.all.short-edge.q25\endcsname{$+$144000.0}
\expandafter\def\csname val@rev.desc.d4.sspro.all.short-edge.q75\endcsname{2160.000}
\expandafter\def\csname cnt@rev.desc.d4.sspro.all.short-edge.q75\endcsname{2{,}160}
\expandafter\def\csname pct@rev.desc.d4.sspro.all.short-edge.q75\endcsname{216000.0}
\expandafter\def\csname ppt@rev.desc.d4.sspro.all.short-edge.q75\endcsname{$+$216000.0}
\expandafter\def\csname val@rev.desc.d4.sspro.all.target-area-fraction.median\endcsname{0.000}
\expandafter\def\csname cnt@rev.desc.d4.sspro.all.target-area-fraction.median\endcsname{0}
\expandafter\def\csname pct@rev.desc.d4.sspro.all.target-area-fraction.median\endcsname{0.0}
\expandafter\def\csname ppt@rev.desc.d4.sspro.all.target-area-fraction.median\endcsname{0.0}
\expandafter\def\csname val@rev.desc.d4.sspro.all.target-area-fraction.n\endcsname{176}
\expandafter\def\csname cnt@rev.desc.d4.sspro.all.target-area-fraction.n\endcsname{176}
\expandafter\def\csname pct@rev.desc.d4.sspro.all.target-area-fraction.n\endcsname{17600.0}
\expandafter\def\csname ppt@rev.desc.d4.sspro.all.target-area-fraction.n\endcsname{$+$17600.0}
\expandafter\def\csname val@rev.desc.d4.sspro.all.target-area-fraction.q25\endcsname{0.000}
\expandafter\def\csname cnt@rev.desc.d4.sspro.all.target-area-fraction.q25\endcsname{0}
\expandafter\def\csname pct@rev.desc.d4.sspro.all.target-area-fraction.q25\endcsname{0.0}
\expandafter\def\csname ppt@rev.desc.d4.sspro.all.target-area-fraction.q25\endcsname{0.0}
\expandafter\def\csname val@rev.desc.d4.sspro.all.target-area-fraction.q75\endcsname{0.001}
\expandafter\def\csname cnt@rev.desc.d4.sspro.all.target-area-fraction.q75\endcsname{0}
\expandafter\def\csname pct@rev.desc.d4.sspro.all.target-area-fraction.q75\endcsname{0.1}
\expandafter\def\csname ppt@rev.desc.d4.sspro.all.target-area-fraction.q75\endcsname{$+$0.1}
\expandafter\def\csname val@rev.desc.d4.sspro.all.target-short-side-direct-view.median\endcsname{8.119}
\expandafter\def\csname cnt@rev.desc.d4.sspro.all.target-short-side-direct-view.median\endcsname{8}
\expandafter\def\csname pct@rev.desc.d4.sspro.all.target-short-side-direct-view.median\endcsname{811.9}
\expandafter\def\csname ppt@rev.desc.d4.sspro.all.target-short-side-direct-view.median\endcsname{$+$811.9}
\expandafter\def\csname val@rev.desc.d4.sspro.all.target-short-side-direct-view.n\endcsname{176}
\expandafter\def\csname cnt@rev.desc.d4.sspro.all.target-short-side-direct-view.n\endcsname{176}
\expandafter\def\csname pct@rev.desc.d4.sspro.all.target-short-side-direct-view.n\endcsname{17600.0}
\expandafter\def\csname ppt@rev.desc.d4.sspro.all.target-short-side-direct-view.n\endcsname{$+$17600.0}
\expandafter\def\csname val@rev.desc.d4.sspro.all.target-short-side-direct-view.q25\endcsname{6.578}
\expandafter\def\csname cnt@rev.desc.d4.sspro.all.target-short-side-direct-view.q25\endcsname{7}
\expandafter\def\csname pct@rev.desc.d4.sspro.all.target-short-side-direct-view.q25\endcsname{657.8}
\expandafter\def\csname ppt@rev.desc.d4.sspro.all.target-short-side-direct-view.q25\endcsname{$+$657.8}
\expandafter\def\csname val@rev.desc.d4.sspro.all.target-short-side-direct-view.q75\endcsname{10.400}
\expandafter\def\csname cnt@rev.desc.d4.sspro.all.target-short-side-direct-view.q75\endcsname{10}
\expandafter\def\csname pct@rev.desc.d4.sspro.all.target-short-side-direct-view.q75\endcsname{1040.0}
\expandafter\def\csname ppt@rev.desc.d4.sspro.all.target-short-side-direct-view.q75\endcsname{$+$1040.0}
\expandafter\def\csname val@rev.desc.d4.sspro.all.target-short-side-native.median\endcsname{23.000}
\expandafter\def\csname cnt@rev.desc.d4.sspro.all.target-short-side-native.median\endcsname{23}
\expandafter\def\csname pct@rev.desc.d4.sspro.all.target-short-side-native.median\endcsname{2300.0}
\expandafter\def\csname ppt@rev.desc.d4.sspro.all.target-short-side-native.median\endcsname{$+$2300.0}
\expandafter\def\csname val@rev.desc.d4.sspro.all.target-short-side-native.n\endcsname{176}
\expandafter\def\csname cnt@rev.desc.d4.sspro.all.target-short-side-native.n\endcsname{176}
\expandafter\def\csname pct@rev.desc.d4.sspro.all.target-short-side-native.n\endcsname{17600.0}
\expandafter\def\csname ppt@rev.desc.d4.sspro.all.target-short-side-native.n\endcsname{$+$17600.0}
\expandafter\def\csname val@rev.desc.d4.sspro.all.target-short-side-native.q25\endcsname{19.000}
\expandafter\def\csname cnt@rev.desc.d4.sspro.all.target-short-side-native.q25\endcsname{19}
\expandafter\def\csname pct@rev.desc.d4.sspro.all.target-short-side-native.q25\endcsname{1900.0}
\expandafter\def\csname ppt@rev.desc.d4.sspro.all.target-short-side-native.q25\endcsname{$+$1900.0}
\expandafter\def\csname val@rev.desc.d4.sspro.all.target-short-side-native.q75\endcsname{30.000}
\expandafter\def\csname cnt@rev.desc.d4.sspro.all.target-short-side-native.q75\endcsname{30}
\expandafter\def\csname pct@rev.desc.d4.sspro.all.target-short-side-native.q75\endcsname{3000.0}
\expandafter\def\csname ppt@rev.desc.d4.sspro.all.target-short-side-native.q75\endcsname{$+$3000.0}
\expandafter\def\csname val@rev.desc.d4.sspro.census-median-target-side\endcsname{8.119}
\expandafter\def\csname cnt@rev.desc.d4.sspro.census-median-target-side\endcsname{8}
\expandafter\def\csname pct@rev.desc.d4.sspro.census-median-target-side\endcsname{811.9}
\expandafter\def\csname ppt@rev.desc.d4.sspro.census-median-target-side\endcsname{$+$811.9}
\expandafter\def\csname val@rev.desc.d4.sspro.reproduces-census-median-target-side\endcsname{1}
\expandafter\def\csname cnt@rev.desc.d4.sspro.reproduces-census-median-target-side\endcsname{1}
\expandafter\def\csname pct@rev.desc.d4.sspro.reproduces-census-median-target-side\endcsname{100.0}
\expandafter\def\csname ppt@rev.desc.d4.sspro.reproduces-census-median-target-side\endcsname{$+$100.0}
\expandafter\def\csname val@rev.desc.d4.uiv.A.agree-threshold-over-target-diag.median\endcsname{0.660}
\expandafter\def\csname cnt@rev.desc.d4.uiv.A.agree-threshold-over-target-diag.median\endcsname{1}
\expandafter\def\csname pct@rev.desc.d4.uiv.A.agree-threshold-over-target-diag.median\endcsname{66.0}
\expandafter\def\csname ppt@rev.desc.d4.uiv.A.agree-threshold-over-target-diag.median\endcsname{$+$66.0}
\expandafter\def\csname val@rev.desc.d4.uiv.A.agree-threshold-over-target-diag.n\endcsname{64}
\expandafter\def\csname cnt@rev.desc.d4.uiv.A.agree-threshold-over-target-diag.n\endcsname{64}
\expandafter\def\csname pct@rev.desc.d4.uiv.A.agree-threshold-over-target-diag.n\endcsname{6400.0}
\expandafter\def\csname ppt@rev.desc.d4.uiv.A.agree-threshold-over-target-diag.n\endcsname{$+$6400.0}
\expandafter\def\csname val@rev.desc.d4.uiv.A.agree-threshold-over-target-diag.q25\endcsname{0.408}
\expandafter\def\csname cnt@rev.desc.d4.uiv.A.agree-threshold-over-target-diag.q25\endcsname{0}
\expandafter\def\csname pct@rev.desc.d4.uiv.A.agree-threshold-over-target-diag.q25\endcsname{40.8}
\expandafter\def\csname ppt@rev.desc.d4.uiv.A.agree-threshold-over-target-diag.q25\endcsname{$+$40.8}
\expandafter\def\csname val@rev.desc.d4.uiv.A.agree-threshold-over-target-diag.q75\endcsname{0.852}
\expandafter\def\csname cnt@rev.desc.d4.uiv.A.agree-threshold-over-target-diag.q75\endcsname{1}
\expandafter\def\csname pct@rev.desc.d4.uiv.A.agree-threshold-over-target-diag.q75\endcsname{85.2}
\expandafter\def\csname ppt@rev.desc.d4.uiv.A.agree-threshold-over-target-diag.q75\endcsname{$+$85.2}
\expandafter\def\csname val@rev.desc.d4.uiv.A.agree-threshold-over-target-short-side.median\endcsname{1.427}
\expandafter\def\csname cnt@rev.desc.d4.uiv.A.agree-threshold-over-target-short-side.median\endcsname{1}
\expandafter\def\csname pct@rev.desc.d4.uiv.A.agree-threshold-over-target-short-side.median\endcsname{142.7}
\expandafter\def\csname ppt@rev.desc.d4.uiv.A.agree-threshold-over-target-short-side.median\endcsname{$+$142.7}
\expandafter\def\csname val@rev.desc.d4.uiv.A.agree-threshold-over-target-short-side.n\endcsname{64}
\expandafter\def\csname cnt@rev.desc.d4.uiv.A.agree-threshold-over-target-short-side.n\endcsname{64}
\expandafter\def\csname pct@rev.desc.d4.uiv.A.agree-threshold-over-target-short-side.n\endcsname{6400.0}
\expandafter\def\csname ppt@rev.desc.d4.uiv.A.agree-threshold-over-target-short-side.n\endcsname{$+$6400.0}
\expandafter\def\csname val@rev.desc.d4.uiv.A.agree-threshold-over-target-short-side.q25\endcsname{1.091}
\expandafter\def\csname cnt@rev.desc.d4.uiv.A.agree-threshold-over-target-short-side.q25\endcsname{1}
\expandafter\def\csname pct@rev.desc.d4.uiv.A.agree-threshold-over-target-short-side.q25\endcsname{109.1}
\expandafter\def\csname ppt@rev.desc.d4.uiv.A.agree-threshold-over-target-short-side.q25\endcsname{$+$109.1}
\expandafter\def\csname val@rev.desc.d4.uiv.A.agree-threshold-over-target-short-side.q75\endcsname{1.672}
\expandafter\def\csname cnt@rev.desc.d4.uiv.A.agree-threshold-over-target-short-side.q75\endcsname{2}
\expandafter\def\csname pct@rev.desc.d4.uiv.A.agree-threshold-over-target-short-side.q75\endcsname{167.2}
\expandafter\def\csname ppt@rev.desc.d4.uiv.A.agree-threshold-over-target-short-side.q75\endcsname{$+$167.2}
\expandafter\def\csname val@rev.desc.d4.uiv.A.direct-view-long-edge.median\endcsname{1024.000}
\expandafter\def\csname cnt@rev.desc.d4.uiv.A.direct-view-long-edge.median\endcsname{1{,}024}
\expandafter\def\csname pct@rev.desc.d4.uiv.A.direct-view-long-edge.median\endcsname{102400.0}
\expandafter\def\csname ppt@rev.desc.d4.uiv.A.direct-view-long-edge.median\endcsname{$+$102400.0}
\expandafter\def\csname val@rev.desc.d4.uiv.A.direct-view-long-edge.n\endcsname{64}
\expandafter\def\csname cnt@rev.desc.d4.uiv.A.direct-view-long-edge.n\endcsname{64}
\expandafter\def\csname pct@rev.desc.d4.uiv.A.direct-view-long-edge.n\endcsname{6400.0}
\expandafter\def\csname ppt@rev.desc.d4.uiv.A.direct-view-long-edge.n\endcsname{$+$6400.0}
\expandafter\def\csname val@rev.desc.d4.uiv.A.direct-view-long-edge.q25\endcsname{1024.000}
\expandafter\def\csname cnt@rev.desc.d4.uiv.A.direct-view-long-edge.q25\endcsname{1{,}024}
\expandafter\def\csname pct@rev.desc.d4.uiv.A.direct-view-long-edge.q25\endcsname{102400.0}
\expandafter\def\csname ppt@rev.desc.d4.uiv.A.direct-view-long-edge.q25\endcsname{$+$102400.0}
\expandafter\def\csname val@rev.desc.d4.uiv.A.direct-view-long-edge.q75\endcsname{1024.000}
\expandafter\def\csname cnt@rev.desc.d4.uiv.A.direct-view-long-edge.q75\endcsname{1{,}024}
\expandafter\def\csname pct@rev.desc.d4.uiv.A.direct-view-long-edge.q75\endcsname{102400.0}
\expandafter\def\csname ppt@rev.desc.d4.uiv.A.direct-view-long-edge.q75\endcsname{$+$102400.0}
\expandafter\def\csname val@rev.desc.d4.uiv.A.downscale-factor.median\endcsname{0.533}
\expandafter\def\csname cnt@rev.desc.d4.uiv.A.downscale-factor.median\endcsname{1}
\expandafter\def\csname pct@rev.desc.d4.uiv.A.downscale-factor.median\endcsname{53.3}
\expandafter\def\csname ppt@rev.desc.d4.uiv.A.downscale-factor.median\endcsname{$+$53.3}
\expandafter\def\csname val@rev.desc.d4.uiv.A.downscale-factor.n\endcsname{64}
\expandafter\def\csname cnt@rev.desc.d4.uiv.A.downscale-factor.n\endcsname{64}
\expandafter\def\csname pct@rev.desc.d4.uiv.A.downscale-factor.n\endcsname{6400.0}
\expandafter\def\csname ppt@rev.desc.d4.uiv.A.downscale-factor.n\endcsname{$+$6400.0}
\expandafter\def\csname val@rev.desc.d4.uiv.A.downscale-factor.q25\endcsname{0.443}
\expandafter\def\csname cnt@rev.desc.d4.uiv.A.downscale-factor.q25\endcsname{0}
\expandafter\def\csname pct@rev.desc.d4.uiv.A.downscale-factor.q25\endcsname{44.3}
\expandafter\def\csname ppt@rev.desc.d4.uiv.A.downscale-factor.q25\endcsname{$+$44.3}
\expandafter\def\csname val@rev.desc.d4.uiv.A.downscale-factor.q75\endcsname{0.610}
\expandafter\def\csname cnt@rev.desc.d4.uiv.A.downscale-factor.q75\endcsname{1}
\expandafter\def\csname pct@rev.desc.d4.uiv.A.downscale-factor.q75\endcsname{61.0}
\expandafter\def\csname ppt@rev.desc.d4.uiv.A.downscale-factor.q75\endcsname{$+$61.0}
\expandafter\def\csname val@rev.desc.d4.uiv.A.long-edge.median\endcsname{1920.000}
\expandafter\def\csname cnt@rev.desc.d4.uiv.A.long-edge.median\endcsname{1{,}920}
\expandafter\def\csname pct@rev.desc.d4.uiv.A.long-edge.median\endcsname{192000.0}
\expandafter\def\csname ppt@rev.desc.d4.uiv.A.long-edge.median\endcsname{$+$192000.0}
\expandafter\def\csname val@rev.desc.d4.uiv.A.long-edge.n\endcsname{64}
\expandafter\def\csname cnt@rev.desc.d4.uiv.A.long-edge.n\endcsname{64}
\expandafter\def\csname pct@rev.desc.d4.uiv.A.long-edge.n\endcsname{6400.0}
\expandafter\def\csname ppt@rev.desc.d4.uiv.A.long-edge.n\endcsname{$+$6400.0}
\expandafter\def\csname val@rev.desc.d4.uiv.A.long-edge.q25\endcsname{1680.000}
\expandafter\def\csname cnt@rev.desc.d4.uiv.A.long-edge.q25\endcsname{1{,}680}
\expandafter\def\csname pct@rev.desc.d4.uiv.A.long-edge.q25\endcsname{168000.0}
\expandafter\def\csname ppt@rev.desc.d4.uiv.A.long-edge.q25\endcsname{$+$168000.0}
\expandafter\def\csname val@rev.desc.d4.uiv.A.long-edge.q75\endcsname{2310.000}
\expandafter\def\csname cnt@rev.desc.d4.uiv.A.long-edge.q75\endcsname{2{,}310}
\expandafter\def\csname pct@rev.desc.d4.uiv.A.long-edge.q75\endcsname{231000.0}
\expandafter\def\csname ppt@rev.desc.d4.uiv.A.long-edge.q75\endcsname{$+$231000.0}
\expandafter\def\csname val@rev.desc.d4.uiv.A.megapixels.median\endcsname{1.974}
\expandafter\def\csname cnt@rev.desc.d4.uiv.A.megapixels.median\endcsname{2}
\expandafter\def\csname pct@rev.desc.d4.uiv.A.megapixels.median\endcsname{197.4}
\expandafter\def\csname ppt@rev.desc.d4.uiv.A.megapixels.median\endcsname{$+$197.4}
\expandafter\def\csname val@rev.desc.d4.uiv.A.megapixels.n\endcsname{64}
\expandafter\def\csname cnt@rev.desc.d4.uiv.A.megapixels.n\endcsname{64}
\expandafter\def\csname pct@rev.desc.d4.uiv.A.megapixels.n\endcsname{6400.0}
\expandafter\def\csname ppt@rev.desc.d4.uiv.A.megapixels.n\endcsname{$+$6400.0}
\expandafter\def\csname val@rev.desc.d4.uiv.A.megapixels.q25\endcsname{1.754}
\expandafter\def\csname cnt@rev.desc.d4.uiv.A.megapixels.q25\endcsname{2}
\expandafter\def\csname pct@rev.desc.d4.uiv.A.megapixels.q25\endcsname{175.4}
\expandafter\def\csname ppt@rev.desc.d4.uiv.A.megapixels.q25\endcsname{$+$175.4}
\expandafter\def\csname val@rev.desc.d4.uiv.A.megapixels.q75\endcsname{3.210}
\expandafter\def\csname cnt@rev.desc.d4.uiv.A.megapixels.q75\endcsname{3}
\expandafter\def\csname pct@rev.desc.d4.uiv.A.megapixels.q75\endcsname{321.0}
\expandafter\def\csname ppt@rev.desc.d4.uiv.A.megapixels.q75\endcsname{$+$321.0}
\expandafter\def\csname val@rev.desc.d4.uiv.A.n-apps\endcsname{8}
\expandafter\def\csname cnt@rev.desc.d4.uiv.A.n-apps\endcsname{8}
\expandafter\def\csname pct@rev.desc.d4.uiv.A.n-apps\endcsname{800.0}
\expandafter\def\csname ppt@rev.desc.d4.uiv.A.n-apps\endcsname{$+$800.0}
\expandafter\def\csname val@rev.desc.d4.uiv.A.n-tasks\endcsname{64}
\expandafter\def\csname cnt@rev.desc.d4.uiv.A.n-tasks\endcsname{64}
\expandafter\def\csname pct@rev.desc.d4.uiv.A.n-tasks\endcsname{6400.0}
\expandafter\def\csname ppt@rev.desc.d4.uiv.A.n-tasks\endcsname{$+$6400.0}
\expandafter\def\csname val@rev.desc.d4.uiv.A.share-downscaled\endcsname{1.000}
\expandafter\def\csname cnt@rev.desc.d4.uiv.A.share-downscaled\endcsname{1}
\expandafter\def\csname pct@rev.desc.d4.uiv.A.share-downscaled\endcsname{100.0}
\expandafter\def\csname ppt@rev.desc.d4.uiv.A.share-downscaled\endcsname{$+$100.0}
\expandafter\def\csname val@rev.desc.d4.uiv.A.share-threshold-gt-target-short-side\endcsname{0.828}
\expandafter\def\csname cnt@rev.desc.d4.uiv.A.share-threshold-gt-target-short-side\endcsname{1}
\expandafter\def\csname pct@rev.desc.d4.uiv.A.share-threshold-gt-target-short-side\endcsname{82.8}
\expandafter\def\csname ppt@rev.desc.d4.uiv.A.share-threshold-gt-target-short-side\endcsname{$+$82.8}
\expandafter\def\csname val@rev.desc.d4.uiv.A.short-edge.median\endcsname{1044.000}
\expandafter\def\csname cnt@rev.desc.d4.uiv.A.short-edge.median\endcsname{1{,}044}
\expandafter\def\csname pct@rev.desc.d4.uiv.A.short-edge.median\endcsname{104400.0}
\expandafter\def\csname ppt@rev.desc.d4.uiv.A.short-edge.median\endcsname{$+$104400.0}
\expandafter\def\csname val@rev.desc.d4.uiv.A.short-edge.n\endcsname{64}
\expandafter\def\csname cnt@rev.desc.d4.uiv.A.short-edge.n\endcsname{64}
\expandafter\def\csname pct@rev.desc.d4.uiv.A.short-edge.n\endcsname{6400.0}
\expandafter\def\csname ppt@rev.desc.d4.uiv.A.short-edge.n\endcsname{$+$6400.0}
\expandafter\def\csname val@rev.desc.d4.uiv.A.short-edge.q25\endcsname{1019.500}
\expandafter\def\csname cnt@rev.desc.d4.uiv.A.short-edge.q25\endcsname{1{,}020}
\expandafter\def\csname pct@rev.desc.d4.uiv.A.short-edge.q25\endcsname{101950.0}
\expandafter\def\csname ppt@rev.desc.d4.uiv.A.short-edge.q25\endcsname{$+$101950.0}
\expandafter\def\csname val@rev.desc.d4.uiv.A.short-edge.q75\endcsname{1289.500}
\expandafter\def\csname cnt@rev.desc.d4.uiv.A.short-edge.q75\endcsname{1{,}290}
\expandafter\def\csname pct@rev.desc.d4.uiv.A.short-edge.q75\endcsname{128950.0}
\expandafter\def\csname ppt@rev.desc.d4.uiv.A.short-edge.q75\endcsname{$+$128950.0}
\expandafter\def\csname val@rev.desc.d4.uiv.A.target-area-fraction.median\endcsname{0.001}
\expandafter\def\csname cnt@rev.desc.d4.uiv.A.target-area-fraction.median\endcsname{0}
\expandafter\def\csname pct@rev.desc.d4.uiv.A.target-area-fraction.median\endcsname{0.1}
\expandafter\def\csname ppt@rev.desc.d4.uiv.A.target-area-fraction.median\endcsname{$+$0.1}
\expandafter\def\csname val@rev.desc.d4.uiv.A.target-area-fraction.n\endcsname{64}
\expandafter\def\csname cnt@rev.desc.d4.uiv.A.target-area-fraction.n\endcsname{64}
\expandafter\def\csname pct@rev.desc.d4.uiv.A.target-area-fraction.n\endcsname{6400.0}
\expandafter\def\csname ppt@rev.desc.d4.uiv.A.target-area-fraction.n\endcsname{$+$6400.0}
\expandafter\def\csname val@rev.desc.d4.uiv.A.target-area-fraction.q25\endcsname{0.001}
\expandafter\def\csname cnt@rev.desc.d4.uiv.A.target-area-fraction.q25\endcsname{0}
\expandafter\def\csname pct@rev.desc.d4.uiv.A.target-area-fraction.q25\endcsname{0.1}
\expandafter\def\csname ppt@rev.desc.d4.uiv.A.target-area-fraction.q25\endcsname{$+$0.1}
\expandafter\def\csname val@rev.desc.d4.uiv.A.target-area-fraction.q75\endcsname{0.002}
\expandafter\def\csname cnt@rev.desc.d4.uiv.A.target-area-fraction.q75\endcsname{0}
\expandafter\def\csname pct@rev.desc.d4.uiv.A.target-area-fraction.q75\endcsname{0.2}
\expandafter\def\csname ppt@rev.desc.d4.uiv.A.target-area-fraction.q75\endcsname{$+$0.2}
\expandafter\def\csname val@rev.desc.d4.uiv.A.target-short-side-direct-view.median\endcsname{16.343}
\expandafter\def\csname cnt@rev.desc.d4.uiv.A.target-short-side-direct-view.median\endcsname{16}
\expandafter\def\csname pct@rev.desc.d4.uiv.A.target-short-side-direct-view.median\endcsname{1634.3}
\expandafter\def\csname ppt@rev.desc.d4.uiv.A.target-short-side-direct-view.median\endcsname{$+$1634.3}
\expandafter\def\csname val@rev.desc.d4.uiv.A.target-short-side-direct-view.n\endcsname{64}
\expandafter\def\csname cnt@rev.desc.d4.uiv.A.target-short-side-direct-view.n\endcsname{64}
\expandafter\def\csname pct@rev.desc.d4.uiv.A.target-short-side-direct-view.n\endcsname{6400.0}
\expandafter\def\csname ppt@rev.desc.d4.uiv.A.target-short-side-direct-view.n\endcsname{$+$6400.0}
\expandafter\def\csname val@rev.desc.d4.uiv.A.target-short-side-direct-view.q25\endcsname{13.979}
\expandafter\def\csname cnt@rev.desc.d4.uiv.A.target-short-side-direct-view.q25\endcsname{14}
\expandafter\def\csname pct@rev.desc.d4.uiv.A.target-short-side-direct-view.q25\endcsname{1397.9}
\expandafter\def\csname ppt@rev.desc.d4.uiv.A.target-short-side-direct-view.q25\endcsname{$+$1397.9}
\expandafter\def\csname val@rev.desc.d4.uiv.A.target-short-side-direct-view.q75\endcsname{21.458}
\expandafter\def\csname cnt@rev.desc.d4.uiv.A.target-short-side-direct-view.q75\endcsname{21}
\expandafter\def\csname pct@rev.desc.d4.uiv.A.target-short-side-direct-view.q75\endcsname{2145.8}
\expandafter\def\csname ppt@rev.desc.d4.uiv.A.target-short-side-direct-view.q75\endcsname{$+$2145.8}
\expandafter\def\csname val@rev.desc.d4.uiv.A.target-short-side-native.median\endcsname{30.849}
\expandafter\def\csname cnt@rev.desc.d4.uiv.A.target-short-side-native.median\endcsname{31}
\expandafter\def\csname pct@rev.desc.d4.uiv.A.target-short-side-native.median\endcsname{3084.9}
\expandafter\def\csname ppt@rev.desc.d4.uiv.A.target-short-side-native.median\endcsname{$+$3084.9}
\expandafter\def\csname val@rev.desc.d4.uiv.A.target-short-side-native.n\endcsname{64}
\expandafter\def\csname cnt@rev.desc.d4.uiv.A.target-short-side-native.n\endcsname{64}
\expandafter\def\csname pct@rev.desc.d4.uiv.A.target-short-side-native.n\endcsname{6400.0}
\expandafter\def\csname ppt@rev.desc.d4.uiv.A.target-short-side-native.n\endcsname{$+$6400.0}
\expandafter\def\csname val@rev.desc.d4.uiv.A.target-short-side-native.q25\endcsname{24.622}
\expandafter\def\csname cnt@rev.desc.d4.uiv.A.target-short-side-native.q25\endcsname{25}
\expandafter\def\csname pct@rev.desc.d4.uiv.A.target-short-side-native.q25\endcsname{2462.2}
\expandafter\def\csname ppt@rev.desc.d4.uiv.A.target-short-side-native.q25\endcsname{$+$2462.2}
\expandafter\def\csname val@rev.desc.d4.uiv.A.target-short-side-native.q75\endcsname{44.314}
\expandafter\def\csname cnt@rev.desc.d4.uiv.A.target-short-side-native.q75\endcsname{44}
\expandafter\def\csname pct@rev.desc.d4.uiv.A.target-short-side-native.q75\endcsname{4431.4}
\expandafter\def\csname ppt@rev.desc.d4.uiv.A.target-short-side-native.q75\endcsname{$+$4431.4}
\expandafter\def\csname val@rev.desc.d4.uiv.C.agree-threshold-over-target-diag.median\endcsname{0.836}
\expandafter\def\csname cnt@rev.desc.d4.uiv.C.agree-threshold-over-target-diag.median\endcsname{1}
\expandafter\def\csname pct@rev.desc.d4.uiv.C.agree-threshold-over-target-diag.median\endcsname{83.6}
\expandafter\def\csname ppt@rev.desc.d4.uiv.C.agree-threshold-over-target-diag.median\endcsname{$+$83.6}
\expandafter\def\csname val@rev.desc.d4.uiv.C.agree-threshold-over-target-diag.n\endcsname{96}
\expandafter\def\csname cnt@rev.desc.d4.uiv.C.agree-threshold-over-target-diag.n\endcsname{96}
\expandafter\def\csname pct@rev.desc.d4.uiv.C.agree-threshold-over-target-diag.n\endcsname{9600.0}
\expandafter\def\csname ppt@rev.desc.d4.uiv.C.agree-threshold-over-target-diag.n\endcsname{$+$9600.0}
\expandafter\def\csname val@rev.desc.d4.uiv.C.agree-threshold-over-target-diag.q25\endcsname{0.638}
\expandafter\def\csname cnt@rev.desc.d4.uiv.C.agree-threshold-over-target-diag.q25\endcsname{1}
\expandafter\def\csname pct@rev.desc.d4.uiv.C.agree-threshold-over-target-diag.q25\endcsname{63.8}
\expandafter\def\csname ppt@rev.desc.d4.uiv.C.agree-threshold-over-target-diag.q25\endcsname{$+$63.8}
\expandafter\def\csname val@rev.desc.d4.uiv.C.agree-threshold-over-target-diag.q75\endcsname{1.110}
\expandafter\def\csname cnt@rev.desc.d4.uiv.C.agree-threshold-over-target-diag.q75\endcsname{1}
\expandafter\def\csname pct@rev.desc.d4.uiv.C.agree-threshold-over-target-diag.q75\endcsname{111.0}
\expandafter\def\csname ppt@rev.desc.d4.uiv.C.agree-threshold-over-target-diag.q75\endcsname{$+$111.0}
\expandafter\def\csname val@rev.desc.d4.uiv.C.agree-threshold-over-target-short-side.median\endcsname{1.562}
\expandafter\def\csname cnt@rev.desc.d4.uiv.C.agree-threshold-over-target-short-side.median\endcsname{2}
\expandafter\def\csname pct@rev.desc.d4.uiv.C.agree-threshold-over-target-short-side.median\endcsname{156.2}
\expandafter\def\csname ppt@rev.desc.d4.uiv.C.agree-threshold-over-target-short-side.median\endcsname{$+$156.2}
\expandafter\def\csname val@rev.desc.d4.uiv.C.agree-threshold-over-target-short-side.n\endcsname{96}
\expandafter\def\csname cnt@rev.desc.d4.uiv.C.agree-threshold-over-target-short-side.n\endcsname{96}
\expandafter\def\csname pct@rev.desc.d4.uiv.C.agree-threshold-over-target-short-side.n\endcsname{9600.0}
\expandafter\def\csname ppt@rev.desc.d4.uiv.C.agree-threshold-over-target-short-side.n\endcsname{$+$9600.0}
\expandafter\def\csname val@rev.desc.d4.uiv.C.agree-threshold-over-target-short-side.q25\endcsname{1.298}
\expandafter\def\csname cnt@rev.desc.d4.uiv.C.agree-threshold-over-target-short-side.q25\endcsname{1}
\expandafter\def\csname pct@rev.desc.d4.uiv.C.agree-threshold-over-target-short-side.q25\endcsname{129.8}
\expandafter\def\csname ppt@rev.desc.d4.uiv.C.agree-threshold-over-target-short-side.q25\endcsname{$+$129.8}
\expandafter\def\csname val@rev.desc.d4.uiv.C.agree-threshold-over-target-short-side.q75\endcsname{1.827}
\expandafter\def\csname cnt@rev.desc.d4.uiv.C.agree-threshold-over-target-short-side.q75\endcsname{2}
\expandafter\def\csname pct@rev.desc.d4.uiv.C.agree-threshold-over-target-short-side.q75\endcsname{182.7}
\expandafter\def\csname ppt@rev.desc.d4.uiv.C.agree-threshold-over-target-short-side.q75\endcsname{$+$182.7}
\expandafter\def\csname val@rev.desc.d4.uiv.C.direct-view-long-edge.median\endcsname{1024.000}
\expandafter\def\csname cnt@rev.desc.d4.uiv.C.direct-view-long-edge.median\endcsname{1{,}024}
\expandafter\def\csname pct@rev.desc.d4.uiv.C.direct-view-long-edge.median\endcsname{102400.0}
\expandafter\def\csname ppt@rev.desc.d4.uiv.C.direct-view-long-edge.median\endcsname{$+$102400.0}
\expandafter\def\csname val@rev.desc.d4.uiv.C.direct-view-long-edge.n\endcsname{96}
\expandafter\def\csname cnt@rev.desc.d4.uiv.C.direct-view-long-edge.n\endcsname{96}
\expandafter\def\csname pct@rev.desc.d4.uiv.C.direct-view-long-edge.n\endcsname{9600.0}
\expandafter\def\csname ppt@rev.desc.d4.uiv.C.direct-view-long-edge.n\endcsname{$+$9600.0}
\expandafter\def\csname val@rev.desc.d4.uiv.C.direct-view-long-edge.q25\endcsname{1024.000}
\expandafter\def\csname cnt@rev.desc.d4.uiv.C.direct-view-long-edge.q25\endcsname{1{,}024}
\expandafter\def\csname pct@rev.desc.d4.uiv.C.direct-view-long-edge.q25\endcsname{102400.0}
\expandafter\def\csname ppt@rev.desc.d4.uiv.C.direct-view-long-edge.q25\endcsname{$+$102400.0}
\expandafter\def\csname val@rev.desc.d4.uiv.C.direct-view-long-edge.q75\endcsname{1024.000}
\expandafter\def\csname cnt@rev.desc.d4.uiv.C.direct-view-long-edge.q75\endcsname{1{,}024}
\expandafter\def\csname pct@rev.desc.d4.uiv.C.direct-view-long-edge.q75\endcsname{102400.0}
\expandafter\def\csname ppt@rev.desc.d4.uiv.C.direct-view-long-edge.q75\endcsname{$+$102400.0}
\expandafter\def\csname val@rev.desc.d4.uiv.C.downscale-factor.median\endcsname{0.533}
\expandafter\def\csname cnt@rev.desc.d4.uiv.C.downscale-factor.median\endcsname{1}
\expandafter\def\csname pct@rev.desc.d4.uiv.C.downscale-factor.median\endcsname{53.3}
\expandafter\def\csname ppt@rev.desc.d4.uiv.C.downscale-factor.median\endcsname{$+$53.3}
\expandafter\def\csname val@rev.desc.d4.uiv.C.downscale-factor.n\endcsname{96}
\expandafter\def\csname cnt@rev.desc.d4.uiv.C.downscale-factor.n\endcsname{96}
\expandafter\def\csname pct@rev.desc.d4.uiv.C.downscale-factor.n\endcsname{9600.0}
\expandafter\def\csname ppt@rev.desc.d4.uiv.C.downscale-factor.n\endcsname{$+$9600.0}
\expandafter\def\csname val@rev.desc.d4.uiv.C.downscale-factor.q25\endcsname{0.533}
\expandafter\def\csname cnt@rev.desc.d4.uiv.C.downscale-factor.q25\endcsname{1}
\expandafter\def\csname pct@rev.desc.d4.uiv.C.downscale-factor.q25\endcsname{53.3}
\expandafter\def\csname ppt@rev.desc.d4.uiv.C.downscale-factor.q25\endcsname{$+$53.3}
\expandafter\def\csname val@rev.desc.d4.uiv.C.downscale-factor.q75\endcsname{0.578}
\expandafter\def\csname cnt@rev.desc.d4.uiv.C.downscale-factor.q75\endcsname{1}
\expandafter\def\csname pct@rev.desc.d4.uiv.C.downscale-factor.q75\endcsname{57.8}
\expandafter\def\csname ppt@rev.desc.d4.uiv.C.downscale-factor.q75\endcsname{$+$57.8}
\expandafter\def\csname val@rev.desc.d4.uiv.C.long-edge.median\endcsname{1920.000}
\expandafter\def\csname cnt@rev.desc.d4.uiv.C.long-edge.median\endcsname{1{,}920}
\expandafter\def\csname pct@rev.desc.d4.uiv.C.long-edge.median\endcsname{192000.0}
\expandafter\def\csname ppt@rev.desc.d4.uiv.C.long-edge.median\endcsname{$+$192000.0}
\expandafter\def\csname val@rev.desc.d4.uiv.C.long-edge.n\endcsname{96}
\expandafter\def\csname cnt@rev.desc.d4.uiv.C.long-edge.n\endcsname{96}
\expandafter\def\csname pct@rev.desc.d4.uiv.C.long-edge.n\endcsname{9600.0}
\expandafter\def\csname ppt@rev.desc.d4.uiv.C.long-edge.n\endcsname{$+$9600.0}
\expandafter\def\csname val@rev.desc.d4.uiv.C.long-edge.q25\endcsname{1773.000}
\expandafter\def\csname cnt@rev.desc.d4.uiv.C.long-edge.q25\endcsname{1{,}773}
\expandafter\def\csname pct@rev.desc.d4.uiv.C.long-edge.q25\endcsname{177300.0}
\expandafter\def\csname ppt@rev.desc.d4.uiv.C.long-edge.q25\endcsname{$+$177300.0}
\expandafter\def\csname val@rev.desc.d4.uiv.C.long-edge.q75\endcsname{1920.000}
\expandafter\def\csname cnt@rev.desc.d4.uiv.C.long-edge.q75\endcsname{1{,}920}
\expandafter\def\csname pct@rev.desc.d4.uiv.C.long-edge.q75\endcsname{192000.0}
\expandafter\def\csname ppt@rev.desc.d4.uiv.C.long-edge.q75\endcsname{$+$192000.0}
\expandafter\def\csname val@rev.desc.d4.uiv.C.megapixels.median\endcsname{1.968}
\expandafter\def\csname cnt@rev.desc.d4.uiv.C.megapixels.median\endcsname{2}
\expandafter\def\csname pct@rev.desc.d4.uiv.C.megapixels.median\endcsname{196.8}
\expandafter\def\csname ppt@rev.desc.d4.uiv.C.megapixels.median\endcsname{$+$196.8}
\expandafter\def\csname val@rev.desc.d4.uiv.C.megapixels.n\endcsname{96}
\expandafter\def\csname cnt@rev.desc.d4.uiv.C.megapixels.n\endcsname{96}
\expandafter\def\csname pct@rev.desc.d4.uiv.C.megapixels.n\endcsname{9600.0}
\expandafter\def\csname ppt@rev.desc.d4.uiv.C.megapixels.n\endcsname{$+$9600.0}
\expandafter\def\csname val@rev.desc.d4.uiv.C.megapixels.q25\endcsname{1.754}
\expandafter\def\csname cnt@rev.desc.d4.uiv.C.megapixels.q25\endcsname{2}
\expandafter\def\csname pct@rev.desc.d4.uiv.C.megapixels.q25\endcsname{175.4}
\expandafter\def\csname ppt@rev.desc.d4.uiv.C.megapixels.q25\endcsname{$+$175.4}
\expandafter\def\csname val@rev.desc.d4.uiv.C.megapixels.q75\endcsname{2.074}
\expandafter\def\csname cnt@rev.desc.d4.uiv.C.megapixels.q75\endcsname{2}
\expandafter\def\csname pct@rev.desc.d4.uiv.C.megapixels.q75\endcsname{207.4}
\expandafter\def\csname ppt@rev.desc.d4.uiv.C.megapixels.q75\endcsname{$+$207.4}
\expandafter\def\csname val@rev.desc.d4.uiv.C.n-apps\endcsname{12}
\expandafter\def\csname cnt@rev.desc.d4.uiv.C.n-apps\endcsname{12}
\expandafter\def\csname pct@rev.desc.d4.uiv.C.n-apps\endcsname{1200.0}
\expandafter\def\csname ppt@rev.desc.d4.uiv.C.n-apps\endcsname{$+$1200.0}
\expandafter\def\csname val@rev.desc.d4.uiv.C.n-tasks\endcsname{96}
\expandafter\def\csname cnt@rev.desc.d4.uiv.C.n-tasks\endcsname{96}
\expandafter\def\csname pct@rev.desc.d4.uiv.C.n-tasks\endcsname{9600.0}
\expandafter\def\csname ppt@rev.desc.d4.uiv.C.n-tasks\endcsname{$+$9600.0}
\expandafter\def\csname val@rev.desc.d4.uiv.C.share-downscaled\endcsname{1.000}
\expandafter\def\csname cnt@rev.desc.d4.uiv.C.share-downscaled\endcsname{1}
\expandafter\def\csname pct@rev.desc.d4.uiv.C.share-downscaled\endcsname{100.0}
\expandafter\def\csname ppt@rev.desc.d4.uiv.C.share-downscaled\endcsname{$+$100.0}
\expandafter\def\csname val@rev.desc.d4.uiv.C.share-threshold-gt-target-short-side\endcsname{0.906}
\expandafter\def\csname cnt@rev.desc.d4.uiv.C.share-threshold-gt-target-short-side\endcsname{1}
\expandafter\def\csname pct@rev.desc.d4.uiv.C.share-threshold-gt-target-short-side\endcsname{90.6}
\expandafter\def\csname ppt@rev.desc.d4.uiv.C.share-threshold-gt-target-short-side\endcsname{$+$90.6}
\expandafter\def\csname val@rev.desc.d4.uiv.C.short-edge.median\endcsname{1044.000}
\expandafter\def\csname cnt@rev.desc.d4.uiv.C.short-edge.median\endcsname{1{,}044}
\expandafter\def\csname pct@rev.desc.d4.uiv.C.short-edge.median\endcsname{104400.0}
\expandafter\def\csname ppt@rev.desc.d4.uiv.C.short-edge.median\endcsname{$+$104400.0}
\expandafter\def\csname val@rev.desc.d4.uiv.C.short-edge.n\endcsname{96}
\expandafter\def\csname cnt@rev.desc.d4.uiv.C.short-edge.n\endcsname{96}
\expandafter\def\csname pct@rev.desc.d4.uiv.C.short-edge.n\endcsname{9600.0}
\expandafter\def\csname ppt@rev.desc.d4.uiv.C.short-edge.n\endcsname{$+$9600.0}
\expandafter\def\csname val@rev.desc.d4.uiv.C.short-edge.q25\endcsname{953.000}
\expandafter\def\csname cnt@rev.desc.d4.uiv.C.short-edge.q25\endcsname{953}
\expandafter\def\csname pct@rev.desc.d4.uiv.C.short-edge.q25\endcsname{95300.0}
\expandafter\def\csname ppt@rev.desc.d4.uiv.C.short-edge.q25\endcsname{$+$95300.0}
\expandafter\def\csname val@rev.desc.d4.uiv.C.short-edge.q75\endcsname{1080.000}
\expandafter\def\csname cnt@rev.desc.d4.uiv.C.short-edge.q75\endcsname{1{,}080}
\expandafter\def\csname pct@rev.desc.d4.uiv.C.short-edge.q75\endcsname{108000.0}
\expandafter\def\csname ppt@rev.desc.d4.uiv.C.short-edge.q75\endcsname{$+$108000.0}
\expandafter\def\csname val@rev.desc.d4.uiv.C.target-area-fraction.median\endcsname{0.001}
\expandafter\def\csname cnt@rev.desc.d4.uiv.C.target-area-fraction.median\endcsname{0}
\expandafter\def\csname pct@rev.desc.d4.uiv.C.target-area-fraction.median\endcsname{0.1}
\expandafter\def\csname ppt@rev.desc.d4.uiv.C.target-area-fraction.median\endcsname{$+$0.1}
\expandafter\def\csname val@rev.desc.d4.uiv.C.target-area-fraction.n\endcsname{96}
\expandafter\def\csname cnt@rev.desc.d4.uiv.C.target-area-fraction.n\endcsname{96}
\expandafter\def\csname pct@rev.desc.d4.uiv.C.target-area-fraction.n\endcsname{9600.0}
\expandafter\def\csname ppt@rev.desc.d4.uiv.C.target-area-fraction.n\endcsname{$+$9600.0}
\expandafter\def\csname val@rev.desc.d4.uiv.C.target-area-fraction.q25\endcsname{0.000}
\expandafter\def\csname cnt@rev.desc.d4.uiv.C.target-area-fraction.q25\endcsname{0}
\expandafter\def\csname pct@rev.desc.d4.uiv.C.target-area-fraction.q25\endcsname{0.0}
\expandafter\def\csname ppt@rev.desc.d4.uiv.C.target-area-fraction.q25\endcsname{0.0}
\expandafter\def\csname val@rev.desc.d4.uiv.C.target-area-fraction.q75\endcsname{0.001}
\expandafter\def\csname cnt@rev.desc.d4.uiv.C.target-area-fraction.q75\endcsname{0}
\expandafter\def\csname pct@rev.desc.d4.uiv.C.target-area-fraction.q75\endcsname{0.1}
\expandafter\def\csname ppt@rev.desc.d4.uiv.C.target-area-fraction.q75\endcsname{$+$0.1}
\expandafter\def\csname val@rev.desc.d4.uiv.C.target-short-side-direct-view.median\endcsname{15.087}
\expandafter\def\csname cnt@rev.desc.d4.uiv.C.target-short-side-direct-view.median\endcsname{15}
\expandafter\def\csname pct@rev.desc.d4.uiv.C.target-short-side-direct-view.median\endcsname{1508.7}
\expandafter\def\csname ppt@rev.desc.d4.uiv.C.target-short-side-direct-view.median\endcsname{$+$1508.7}
\expandafter\def\csname val@rev.desc.d4.uiv.C.target-short-side-direct-view.n\endcsname{96}
\expandafter\def\csname cnt@rev.desc.d4.uiv.C.target-short-side-direct-view.n\endcsname{96}
\expandafter\def\csname pct@rev.desc.d4.uiv.C.target-short-side-direct-view.n\endcsname{9600.0}
\expandafter\def\csname ppt@rev.desc.d4.uiv.C.target-short-side-direct-view.n\endcsname{$+$9600.0}
\expandafter\def\csname val@rev.desc.d4.uiv.C.target-short-side-direct-view.q25\endcsname{12.800}
\expandafter\def\csname cnt@rev.desc.d4.uiv.C.target-short-side-direct-view.q25\endcsname{13}
\expandafter\def\csname pct@rev.desc.d4.uiv.C.target-short-side-direct-view.q25\endcsname{1280.0}
\expandafter\def\csname ppt@rev.desc.d4.uiv.C.target-short-side-direct-view.q25\endcsname{$+$1280.0}
\expandafter\def\csname val@rev.desc.d4.uiv.C.target-short-side-direct-view.q75\endcsname{17.931}
\expandafter\def\csname cnt@rev.desc.d4.uiv.C.target-short-side-direct-view.q75\endcsname{18}
\expandafter\def\csname pct@rev.desc.d4.uiv.C.target-short-side-direct-view.q75\endcsname{1793.1}
\expandafter\def\csname ppt@rev.desc.d4.uiv.C.target-short-side-direct-view.q75\endcsname{$+$1793.1}
\expandafter\def\csname val@rev.desc.d4.uiv.C.target-short-side-native.median\endcsname{26.596}
\expandafter\def\csname cnt@rev.desc.d4.uiv.C.target-short-side-native.median\endcsname{27}
\expandafter\def\csname pct@rev.desc.d4.uiv.C.target-short-side-native.median\endcsname{2659.6}
\expandafter\def\csname ppt@rev.desc.d4.uiv.C.target-short-side-native.median\endcsname{$+$2659.6}
\expandafter\def\csname val@rev.desc.d4.uiv.C.target-short-side-native.n\endcsname{96}
\expandafter\def\csname cnt@rev.desc.d4.uiv.C.target-short-side-native.n\endcsname{96}
\expandafter\def\csname pct@rev.desc.d4.uiv.C.target-short-side-native.n\endcsname{9600.0}
\expandafter\def\csname ppt@rev.desc.d4.uiv.C.target-short-side-native.n\endcsname{$+$9600.0}
\expandafter\def\csname val@rev.desc.d4.uiv.C.target-short-side-native.q25\endcsname{23.360}
\expandafter\def\csname cnt@rev.desc.d4.uiv.C.target-short-side-native.q25\endcsname{23}
\expandafter\def\csname pct@rev.desc.d4.uiv.C.target-short-side-native.q25\endcsname{2336.0}
\expandafter\def\csname ppt@rev.desc.d4.uiv.C.target-short-side-native.q25\endcsname{$+$2336.0}
\expandafter\def\csname val@rev.desc.d4.uiv.C.target-short-side-native.q75\endcsname{34.217}
\expandafter\def\csname cnt@rev.desc.d4.uiv.C.target-short-side-native.q75\endcsname{34}
\expandafter\def\csname pct@rev.desc.d4.uiv.C.target-short-side-native.q75\endcsname{3421.7}
\expandafter\def\csname ppt@rev.desc.d4.uiv.C.target-short-side-native.q75\endcsname{$+$3421.7}
\expandafter\def\csname val@rev.desc.d4.uiv.T.agree-threshold-over-target-diag.median\endcsname{0.829}
\expandafter\def\csname cnt@rev.desc.d4.uiv.T.agree-threshold-over-target-diag.median\endcsname{1}
\expandafter\def\csname pct@rev.desc.d4.uiv.T.agree-threshold-over-target-diag.median\endcsname{82.9}
\expandafter\def\csname ppt@rev.desc.d4.uiv.T.agree-threshold-over-target-diag.median\endcsname{$+$82.9}
\expandafter\def\csname val@rev.desc.d4.uiv.T.agree-threshold-over-target-diag.n\endcsname{144}
\expandafter\def\csname cnt@rev.desc.d4.uiv.T.agree-threshold-over-target-diag.n\endcsname{144}
\expandafter\def\csname pct@rev.desc.d4.uiv.T.agree-threshold-over-target-diag.n\endcsname{14400.0}
\expandafter\def\csname ppt@rev.desc.d4.uiv.T.agree-threshold-over-target-diag.n\endcsname{$+$14400.0}
\expandafter\def\csname val@rev.desc.d4.uiv.T.agree-threshold-over-target-diag.q25\endcsname{0.610}
\expandafter\def\csname cnt@rev.desc.d4.uiv.T.agree-threshold-over-target-diag.q25\endcsname{1}
\expandafter\def\csname pct@rev.desc.d4.uiv.T.agree-threshold-over-target-diag.q25\endcsname{61.0}
\expandafter\def\csname ppt@rev.desc.d4.uiv.T.agree-threshold-over-target-diag.q25\endcsname{$+$61.0}
\expandafter\def\csname val@rev.desc.d4.uiv.T.agree-threshold-over-target-diag.q75\endcsname{1.069}
\expandafter\def\csname cnt@rev.desc.d4.uiv.T.agree-threshold-over-target-diag.q75\endcsname{1}
\expandafter\def\csname pct@rev.desc.d4.uiv.T.agree-threshold-over-target-diag.q75\endcsname{106.9}
\expandafter\def\csname ppt@rev.desc.d4.uiv.T.agree-threshold-over-target-diag.q75\endcsname{$+$106.9}
\expandafter\def\csname val@rev.desc.d4.uiv.T.agree-threshold-over-target-short-side.median\endcsname{1.478}
\expandafter\def\csname cnt@rev.desc.d4.uiv.T.agree-threshold-over-target-short-side.median\endcsname{1}
\expandafter\def\csname pct@rev.desc.d4.uiv.T.agree-threshold-over-target-short-side.median\endcsname{147.8}
\expandafter\def\csname ppt@rev.desc.d4.uiv.T.agree-threshold-over-target-short-side.median\endcsname{$+$147.8}
\expandafter\def\csname val@rev.desc.d4.uiv.T.agree-threshold-over-target-short-side.n\endcsname{144}
\expandafter\def\csname cnt@rev.desc.d4.uiv.T.agree-threshold-over-target-short-side.n\endcsname{144}
\expandafter\def\csname pct@rev.desc.d4.uiv.T.agree-threshold-over-target-short-side.n\endcsname{14400.0}
\expandafter\def\csname ppt@rev.desc.d4.uiv.T.agree-threshold-over-target-short-side.n\endcsname{$+$14400.0}
\expandafter\def\csname val@rev.desc.d4.uiv.T.agree-threshold-over-target-short-side.q25\endcsname{1.214}
\expandafter\def\csname cnt@rev.desc.d4.uiv.T.agree-threshold-over-target-short-side.q25\endcsname{1}
\expandafter\def\csname pct@rev.desc.d4.uiv.T.agree-threshold-over-target-short-side.q25\endcsname{121.4}
\expandafter\def\csname ppt@rev.desc.d4.uiv.T.agree-threshold-over-target-short-side.q25\endcsname{$+$121.4}
\expandafter\def\csname val@rev.desc.d4.uiv.T.agree-threshold-over-target-short-side.q75\endcsname{1.777}
\expandafter\def\csname cnt@rev.desc.d4.uiv.T.agree-threshold-over-target-short-side.q75\endcsname{2}
\expandafter\def\csname pct@rev.desc.d4.uiv.T.agree-threshold-over-target-short-side.q75\endcsname{177.7}
\expandafter\def\csname ppt@rev.desc.d4.uiv.T.agree-threshold-over-target-short-side.q75\endcsname{$+$177.7}
\expandafter\def\csname val@rev.desc.d4.uiv.T.direct-view-long-edge.median\endcsname{1024.000}
\expandafter\def\csname cnt@rev.desc.d4.uiv.T.direct-view-long-edge.median\endcsname{1{,}024}
\expandafter\def\csname pct@rev.desc.d4.uiv.T.direct-view-long-edge.median\endcsname{102400.0}
\expandafter\def\csname ppt@rev.desc.d4.uiv.T.direct-view-long-edge.median\endcsname{$+$102400.0}
\expandafter\def\csname val@rev.desc.d4.uiv.T.direct-view-long-edge.n\endcsname{144}
\expandafter\def\csname cnt@rev.desc.d4.uiv.T.direct-view-long-edge.n\endcsname{144}
\expandafter\def\csname pct@rev.desc.d4.uiv.T.direct-view-long-edge.n\endcsname{14400.0}
\expandafter\def\csname ppt@rev.desc.d4.uiv.T.direct-view-long-edge.n\endcsname{$+$14400.0}
\expandafter\def\csname val@rev.desc.d4.uiv.T.direct-view-long-edge.q25\endcsname{1024.000}
\expandafter\def\csname cnt@rev.desc.d4.uiv.T.direct-view-long-edge.q25\endcsname{1{,}024}
\expandafter\def\csname pct@rev.desc.d4.uiv.T.direct-view-long-edge.q25\endcsname{102400.0}
\expandafter\def\csname ppt@rev.desc.d4.uiv.T.direct-view-long-edge.q25\endcsname{$+$102400.0}
\expandafter\def\csname val@rev.desc.d4.uiv.T.direct-view-long-edge.q75\endcsname{1024.000}
\expandafter\def\csname cnt@rev.desc.d4.uiv.T.direct-view-long-edge.q75\endcsname{1{,}024}
\expandafter\def\csname pct@rev.desc.d4.uiv.T.direct-view-long-edge.q75\endcsname{102400.0}
\expandafter\def\csname ppt@rev.desc.d4.uiv.T.direct-view-long-edge.q75\endcsname{$+$102400.0}
\expandafter\def\csname val@rev.desc.d4.uiv.T.downscale-factor.median\endcsname{0.533}
\expandafter\def\csname cnt@rev.desc.d4.uiv.T.downscale-factor.median\endcsname{1}
\expandafter\def\csname pct@rev.desc.d4.uiv.T.downscale-factor.median\endcsname{53.3}
\expandafter\def\csname ppt@rev.desc.d4.uiv.T.downscale-factor.median\endcsname{$+$53.3}
\expandafter\def\csname val@rev.desc.d4.uiv.T.downscale-factor.n\endcsname{144}
\expandafter\def\csname cnt@rev.desc.d4.uiv.T.downscale-factor.n\endcsname{144}
\expandafter\def\csname pct@rev.desc.d4.uiv.T.downscale-factor.n\endcsname{14400.0}
\expandafter\def\csname ppt@rev.desc.d4.uiv.T.downscale-factor.n\endcsname{$+$14400.0}
\expandafter\def\csname val@rev.desc.d4.uiv.T.downscale-factor.q25\endcsname{0.533}
\expandafter\def\csname cnt@rev.desc.d4.uiv.T.downscale-factor.q25\endcsname{1}
\expandafter\def\csname pct@rev.desc.d4.uiv.T.downscale-factor.q25\endcsname{53.3}
\expandafter\def\csname ppt@rev.desc.d4.uiv.T.downscale-factor.q25\endcsname{$+$53.3}
\expandafter\def\csname val@rev.desc.d4.uiv.T.downscale-factor.q75\endcsname{0.610}
\expandafter\def\csname cnt@rev.desc.d4.uiv.T.downscale-factor.q75\endcsname{1}
\expandafter\def\csname pct@rev.desc.d4.uiv.T.downscale-factor.q75\endcsname{61.0}
\expandafter\def\csname ppt@rev.desc.d4.uiv.T.downscale-factor.q75\endcsname{$+$61.0}
\expandafter\def\csname val@rev.desc.d4.uiv.T.long-edge.median\endcsname{1920.000}
\expandafter\def\csname cnt@rev.desc.d4.uiv.T.long-edge.median\endcsname{1{,}920}
\expandafter\def\csname pct@rev.desc.d4.uiv.T.long-edge.median\endcsname{192000.0}
\expandafter\def\csname ppt@rev.desc.d4.uiv.T.long-edge.median\endcsname{$+$192000.0}
\expandafter\def\csname val@rev.desc.d4.uiv.T.long-edge.n\endcsname{144}
\expandafter\def\csname cnt@rev.desc.d4.uiv.T.long-edge.n\endcsname{144}
\expandafter\def\csname pct@rev.desc.d4.uiv.T.long-edge.n\endcsname{14400.0}
\expandafter\def\csname ppt@rev.desc.d4.uiv.T.long-edge.n\endcsname{$+$14400.0}
\expandafter\def\csname val@rev.desc.d4.uiv.T.long-edge.q25\endcsname{1680.000}
\expandafter\def\csname cnt@rev.desc.d4.uiv.T.long-edge.q25\endcsname{1{,}680}
\expandafter\def\csname pct@rev.desc.d4.uiv.T.long-edge.q25\endcsname{168000.0}
\expandafter\def\csname ppt@rev.desc.d4.uiv.T.long-edge.q25\endcsname{$+$168000.0}
\expandafter\def\csname val@rev.desc.d4.uiv.T.long-edge.q75\endcsname{1920.000}
\expandafter\def\csname cnt@rev.desc.d4.uiv.T.long-edge.q75\endcsname{1{,}920}
\expandafter\def\csname pct@rev.desc.d4.uiv.T.long-edge.q75\endcsname{192000.0}
\expandafter\def\csname ppt@rev.desc.d4.uiv.T.long-edge.q75\endcsname{$+$192000.0}
\expandafter\def\csname val@rev.desc.d4.uiv.T.megapixels.median\endcsname{1.978}
\expandafter\def\csname cnt@rev.desc.d4.uiv.T.megapixels.median\endcsname{2}
\expandafter\def\csname pct@rev.desc.d4.uiv.T.megapixels.median\endcsname{197.8}
\expandafter\def\csname ppt@rev.desc.d4.uiv.T.megapixels.median\endcsname{$+$197.8}
\expandafter\def\csname val@rev.desc.d4.uiv.T.megapixels.n\endcsname{144}
\expandafter\def\csname cnt@rev.desc.d4.uiv.T.megapixels.n\endcsname{144}
\expandafter\def\csname pct@rev.desc.d4.uiv.T.megapixels.n\endcsname{14400.0}
\expandafter\def\csname ppt@rev.desc.d4.uiv.T.megapixels.n\endcsname{$+$14400.0}
\expandafter\def\csname val@rev.desc.d4.uiv.T.megapixels.q25\endcsname{1.754}
\expandafter\def\csname cnt@rev.desc.d4.uiv.T.megapixels.q25\endcsname{2}
\expandafter\def\csname pct@rev.desc.d4.uiv.T.megapixels.q25\endcsname{175.4}
\expandafter\def\csname ppt@rev.desc.d4.uiv.T.megapixels.q25\endcsname{$+$175.4}
\expandafter\def\csname val@rev.desc.d4.uiv.T.megapixels.q75\endcsname{2.074}
\expandafter\def\csname cnt@rev.desc.d4.uiv.T.megapixels.q75\endcsname{2}
\expandafter\def\csname pct@rev.desc.d4.uiv.T.megapixels.q75\endcsname{207.4}
\expandafter\def\csname ppt@rev.desc.d4.uiv.T.megapixels.q75\endcsname{$+$207.4}
\expandafter\def\csname val@rev.desc.d4.uiv.T.n-apps\endcsname{24}
\expandafter\def\csname cnt@rev.desc.d4.uiv.T.n-apps\endcsname{24}
\expandafter\def\csname pct@rev.desc.d4.uiv.T.n-apps\endcsname{2400.0}
\expandafter\def\csname ppt@rev.desc.d4.uiv.T.n-apps\endcsname{$+$2400.0}
\expandafter\def\csname val@rev.desc.d4.uiv.T.n-tasks\endcsname{144}
\expandafter\def\csname cnt@rev.desc.d4.uiv.T.n-tasks\endcsname{144}
\expandafter\def\csname pct@rev.desc.d4.uiv.T.n-tasks\endcsname{14400.0}
\expandafter\def\csname ppt@rev.desc.d4.uiv.T.n-tasks\endcsname{$+$14400.0}
\expandafter\def\csname val@rev.desc.d4.uiv.T.share-downscaled\endcsname{0.993}
\expandafter\def\csname cnt@rev.desc.d4.uiv.T.share-downscaled\endcsname{1}
\expandafter\def\csname pct@rev.desc.d4.uiv.T.share-downscaled\endcsname{99.3}
\expandafter\def\csname ppt@rev.desc.d4.uiv.T.share-downscaled\endcsname{$+$99.3}
\expandafter\def\csname val@rev.desc.d4.uiv.T.share-threshold-gt-target-short-side\endcsname{0.924}
\expandafter\def\csname cnt@rev.desc.d4.uiv.T.share-threshold-gt-target-short-side\endcsname{1}
\expandafter\def\csname pct@rev.desc.d4.uiv.T.share-threshold-gt-target-short-side\endcsname{92.4}
\expandafter\def\csname ppt@rev.desc.d4.uiv.T.share-threshold-gt-target-short-side\endcsname{$+$92.4}
\expandafter\def\csname val@rev.desc.d4.uiv.T.short-edge.median\endcsname{1044.000}
\expandafter\def\csname cnt@rev.desc.d4.uiv.T.short-edge.median\endcsname{1{,}044}
\expandafter\def\csname pct@rev.desc.d4.uiv.T.short-edge.median\endcsname{104400.0}
\expandafter\def\csname ppt@rev.desc.d4.uiv.T.short-edge.median\endcsname{$+$104400.0}
\expandafter\def\csname val@rev.desc.d4.uiv.T.short-edge.n\endcsname{144}
\expandafter\def\csname cnt@rev.desc.d4.uiv.T.short-edge.n\endcsname{144}
\expandafter\def\csname pct@rev.desc.d4.uiv.T.short-edge.n\endcsname{14400.0}
\expandafter\def\csname ppt@rev.desc.d4.uiv.T.short-edge.n\endcsname{$+$14400.0}
\expandafter\def\csname val@rev.desc.d4.uiv.T.short-edge.q25\endcsname{1008.000}
\expandafter\def\csname cnt@rev.desc.d4.uiv.T.short-edge.q25\endcsname{1{,}008}
\expandafter\def\csname pct@rev.desc.d4.uiv.T.short-edge.q25\endcsname{100800.0}
\expandafter\def\csname ppt@rev.desc.d4.uiv.T.short-edge.q25\endcsname{$+$100800.0}
\expandafter\def\csname val@rev.desc.d4.uiv.T.short-edge.q75\endcsname{1080.000}
\expandafter\def\csname cnt@rev.desc.d4.uiv.T.short-edge.q75\endcsname{1{,}080}
\expandafter\def\csname pct@rev.desc.d4.uiv.T.short-edge.q75\endcsname{108000.0}
\expandafter\def\csname ppt@rev.desc.d4.uiv.T.short-edge.q75\endcsname{$+$108000.0}
\expandafter\def\csname val@rev.desc.d4.uiv.T.target-area-fraction.median\endcsname{0.001}
\expandafter\def\csname cnt@rev.desc.d4.uiv.T.target-area-fraction.median\endcsname{0}
\expandafter\def\csname pct@rev.desc.d4.uiv.T.target-area-fraction.median\endcsname{0.1}
\expandafter\def\csname ppt@rev.desc.d4.uiv.T.target-area-fraction.median\endcsname{$+$0.1}
\expandafter\def\csname val@rev.desc.d4.uiv.T.target-area-fraction.n\endcsname{144}
\expandafter\def\csname cnt@rev.desc.d4.uiv.T.target-area-fraction.n\endcsname{144}
\expandafter\def\csname pct@rev.desc.d4.uiv.T.target-area-fraction.n\endcsname{14400.0}
\expandafter\def\csname ppt@rev.desc.d4.uiv.T.target-area-fraction.n\endcsname{$+$14400.0}
\expandafter\def\csname val@rev.desc.d4.uiv.T.target-area-fraction.q25\endcsname{0.000}
\expandafter\def\csname cnt@rev.desc.d4.uiv.T.target-area-fraction.q25\endcsname{0}
\expandafter\def\csname pct@rev.desc.d4.uiv.T.target-area-fraction.q25\endcsname{0.0}
\expandafter\def\csname ppt@rev.desc.d4.uiv.T.target-area-fraction.q25\endcsname{0.0}
\expandafter\def\csname val@rev.desc.d4.uiv.T.target-area-fraction.q75\endcsname{0.001}
\expandafter\def\csname cnt@rev.desc.d4.uiv.T.target-area-fraction.q75\endcsname{0}
\expandafter\def\csname pct@rev.desc.d4.uiv.T.target-area-fraction.q75\endcsname{0.1}
\expandafter\def\csname ppt@rev.desc.d4.uiv.T.target-area-fraction.q75\endcsname{$+$0.1}
\expandafter\def\csname val@rev.desc.d4.uiv.T.target-short-side-direct-view.median\endcsname{16.093}
\expandafter\def\csname cnt@rev.desc.d4.uiv.T.target-short-side-direct-view.median\endcsname{16}
\expandafter\def\csname pct@rev.desc.d4.uiv.T.target-short-side-direct-view.median\endcsname{1609.3}
\expandafter\def\csname ppt@rev.desc.d4.uiv.T.target-short-side-direct-view.median\endcsname{$+$1609.3}
\expandafter\def\csname val@rev.desc.d4.uiv.T.target-short-side-direct-view.n\endcsname{144}
\expandafter\def\csname cnt@rev.desc.d4.uiv.T.target-short-side-direct-view.n\endcsname{144}
\expandafter\def\csname pct@rev.desc.d4.uiv.T.target-short-side-direct-view.n\endcsname{14400.0}
\expandafter\def\csname ppt@rev.desc.d4.uiv.T.target-short-side-direct-view.n\endcsname{$+$14400.0}
\expandafter\def\csname val@rev.desc.d4.uiv.T.target-short-side-direct-view.q25\endcsname{13.106}
\expandafter\def\csname cnt@rev.desc.d4.uiv.T.target-short-side-direct-view.q25\endcsname{13}
\expandafter\def\csname pct@rev.desc.d4.uiv.T.target-short-side-direct-view.q25\endcsname{1310.6}
\expandafter\def\csname ppt@rev.desc.d4.uiv.T.target-short-side-direct-view.q25\endcsname{$+$1310.6}
\expandafter\def\csname val@rev.desc.d4.uiv.T.target-short-side-direct-view.q75\endcsname{19.206}
\expandafter\def\csname cnt@rev.desc.d4.uiv.T.target-short-side-direct-view.q75\endcsname{19}
\expandafter\def\csname pct@rev.desc.d4.uiv.T.target-short-side-direct-view.q75\endcsname{1920.6}
\expandafter\def\csname ppt@rev.desc.d4.uiv.T.target-short-side-direct-view.q75\endcsname{$+$1920.6}
\expandafter\def\csname val@rev.desc.d4.uiv.T.target-short-side-native.median\endcsname{27.593}
\expandafter\def\csname cnt@rev.desc.d4.uiv.T.target-short-side-native.median\endcsname{28}
\expandafter\def\csname pct@rev.desc.d4.uiv.T.target-short-side-native.median\endcsname{2759.3}
\expandafter\def\csname ppt@rev.desc.d4.uiv.T.target-short-side-native.median\endcsname{$+$2759.3}
\expandafter\def\csname val@rev.desc.d4.uiv.T.target-short-side-native.n\endcsname{144}
\expandafter\def\csname cnt@rev.desc.d4.uiv.T.target-short-side-native.n\endcsname{144}
\expandafter\def\csname pct@rev.desc.d4.uiv.T.target-short-side-native.n\endcsname{14400.0}
\expandafter\def\csname ppt@rev.desc.d4.uiv.T.target-short-side-native.n\endcsname{$+$14400.0}
\expandafter\def\csname val@rev.desc.d4.uiv.T.target-short-side-native.q25\endcsname{22.919}
\expandafter\def\csname cnt@rev.desc.d4.uiv.T.target-short-side-native.q25\endcsname{23}
\expandafter\def\csname pct@rev.desc.d4.uiv.T.target-short-side-native.q25\endcsname{2291.9}
\expandafter\def\csname ppt@rev.desc.d4.uiv.T.target-short-side-native.q25\endcsname{$+$2291.9}
\expandafter\def\csname val@rev.desc.d4.uiv.T.target-short-side-native.q75\endcsname{36.000}
\expandafter\def\csname cnt@rev.desc.d4.uiv.T.target-short-side-native.q75\endcsname{36}
\expandafter\def\csname pct@rev.desc.d4.uiv.T.target-short-side-native.q75\endcsname{3600.0}
\expandafter\def\csname ppt@rev.desc.d4.uiv.T.target-short-side-native.q75\endcsname{$+$3600.0}
\expandafter\def\csname val@rev.desc.d4.uiv.all.agree-threshold-over-target-diag.median\endcsname{0.802}
\expandafter\def\csname cnt@rev.desc.d4.uiv.all.agree-threshold-over-target-diag.median\endcsname{1}
\expandafter\def\csname pct@rev.desc.d4.uiv.all.agree-threshold-over-target-diag.median\endcsname{80.2}
\expandafter\def\csname ppt@rev.desc.d4.uiv.all.agree-threshold-over-target-diag.median\endcsname{$+$80.2}
\expandafter\def\csname val@rev.desc.d4.uiv.all.agree-threshold-over-target-diag.n\endcsname{304}
\expandafter\def\csname cnt@rev.desc.d4.uiv.all.agree-threshold-over-target-diag.n\endcsname{304}
\expandafter\def\csname pct@rev.desc.d4.uiv.all.agree-threshold-over-target-diag.n\endcsname{30400.0}
\expandafter\def\csname ppt@rev.desc.d4.uiv.all.agree-threshold-over-target-diag.n\endcsname{$+$30400.0}
\expandafter\def\csname val@rev.desc.d4.uiv.all.agree-threshold-over-target-diag.q25\endcsname{0.559}
\expandafter\def\csname cnt@rev.desc.d4.uiv.all.agree-threshold-over-target-diag.q25\endcsname{1}
\expandafter\def\csname pct@rev.desc.d4.uiv.all.agree-threshold-over-target-diag.q25\endcsname{55.9}
\expandafter\def\csname ppt@rev.desc.d4.uiv.all.agree-threshold-over-target-diag.q25\endcsname{$+$55.9}
\expandafter\def\csname val@rev.desc.d4.uiv.all.agree-threshold-over-target-diag.q75\endcsname{1.064}
\expandafter\def\csname cnt@rev.desc.d4.uiv.all.agree-threshold-over-target-diag.q75\endcsname{1}
\expandafter\def\csname pct@rev.desc.d4.uiv.all.agree-threshold-over-target-diag.q75\endcsname{106.4}
\expandafter\def\csname ppt@rev.desc.d4.uiv.all.agree-threshold-over-target-diag.q75\endcsname{$+$106.4}
\expandafter\def\csname val@rev.desc.d4.uiv.all.agree-threshold-over-target-short-side.median\endcsname{1.488}
\expandafter\def\csname cnt@rev.desc.d4.uiv.all.agree-threshold-over-target-short-side.median\endcsname{1}
\expandafter\def\csname pct@rev.desc.d4.uiv.all.agree-threshold-over-target-short-side.median\endcsname{148.8}
\expandafter\def\csname ppt@rev.desc.d4.uiv.all.agree-threshold-over-target-short-side.median\endcsname{$+$148.8}
\expandafter\def\csname val@rev.desc.d4.uiv.all.agree-threshold-over-target-short-side.n\endcsname{304}
\expandafter\def\csname cnt@rev.desc.d4.uiv.all.agree-threshold-over-target-short-side.n\endcsname{304}
\expandafter\def\csname pct@rev.desc.d4.uiv.all.agree-threshold-over-target-short-side.n\endcsname{30400.0}
\expandafter\def\csname ppt@rev.desc.d4.uiv.all.agree-threshold-over-target-short-side.n\endcsname{$+$30400.0}
\expandafter\def\csname val@rev.desc.d4.uiv.all.agree-threshold-over-target-short-side.q25\endcsname{1.205}
\expandafter\def\csname cnt@rev.desc.d4.uiv.all.agree-threshold-over-target-short-side.q25\endcsname{1}
\expandafter\def\csname pct@rev.desc.d4.uiv.all.agree-threshold-over-target-short-side.q25\endcsname{120.5}
\expandafter\def\csname ppt@rev.desc.d4.uiv.all.agree-threshold-over-target-short-side.q25\endcsname{$+$120.5}
\expandafter\def\csname val@rev.desc.d4.uiv.all.agree-threshold-over-target-short-side.q75\endcsname{1.785}
\expandafter\def\csname cnt@rev.desc.d4.uiv.all.agree-threshold-over-target-short-side.q75\endcsname{2}
\expandafter\def\csname pct@rev.desc.d4.uiv.all.agree-threshold-over-target-short-side.q75\endcsname{178.5}
\expandafter\def\csname ppt@rev.desc.d4.uiv.all.agree-threshold-over-target-short-side.q75\endcsname{$+$178.5}
\expandafter\def\csname val@rev.desc.d4.uiv.all.direct-view-long-edge.median\endcsname{1024.000}
\expandafter\def\csname cnt@rev.desc.d4.uiv.all.direct-view-long-edge.median\endcsname{1{,}024}
\expandafter\def\csname pct@rev.desc.d4.uiv.all.direct-view-long-edge.median\endcsname{102400.0}
\expandafter\def\csname ppt@rev.desc.d4.uiv.all.direct-view-long-edge.median\endcsname{$+$102400.0}
\expandafter\def\csname val@rev.desc.d4.uiv.all.direct-view-long-edge.n\endcsname{304}
\expandafter\def\csname cnt@rev.desc.d4.uiv.all.direct-view-long-edge.n\endcsname{304}
\expandafter\def\csname pct@rev.desc.d4.uiv.all.direct-view-long-edge.n\endcsname{30400.0}
\expandafter\def\csname ppt@rev.desc.d4.uiv.all.direct-view-long-edge.n\endcsname{$+$30400.0}
\expandafter\def\csname val@rev.desc.d4.uiv.all.direct-view-long-edge.q25\endcsname{1024.000}
\expandafter\def\csname cnt@rev.desc.d4.uiv.all.direct-view-long-edge.q25\endcsname{1{,}024}
\expandafter\def\csname pct@rev.desc.d4.uiv.all.direct-view-long-edge.q25\endcsname{102400.0}
\expandafter\def\csname ppt@rev.desc.d4.uiv.all.direct-view-long-edge.q25\endcsname{$+$102400.0}
\expandafter\def\csname val@rev.desc.d4.uiv.all.direct-view-long-edge.q75\endcsname{1024.000}
\expandafter\def\csname cnt@rev.desc.d4.uiv.all.direct-view-long-edge.q75\endcsname{1{,}024}
\expandafter\def\csname pct@rev.desc.d4.uiv.all.direct-view-long-edge.q75\endcsname{102400.0}
\expandafter\def\csname ppt@rev.desc.d4.uiv.all.direct-view-long-edge.q75\endcsname{$+$102400.0}
\expandafter\def\csname val@rev.desc.d4.uiv.all.downscale-factor.median\endcsname{0.533}
\expandafter\def\csname cnt@rev.desc.d4.uiv.all.downscale-factor.median\endcsname{1}
\expandafter\def\csname pct@rev.desc.d4.uiv.all.downscale-factor.median\endcsname{53.3}
\expandafter\def\csname ppt@rev.desc.d4.uiv.all.downscale-factor.median\endcsname{$+$53.3}
\expandafter\def\csname val@rev.desc.d4.uiv.all.downscale-factor.n\endcsname{304}
\expandafter\def\csname cnt@rev.desc.d4.uiv.all.downscale-factor.n\endcsname{304}
\expandafter\def\csname pct@rev.desc.d4.uiv.all.downscale-factor.n\endcsname{30400.0}
\expandafter\def\csname ppt@rev.desc.d4.uiv.all.downscale-factor.n\endcsname{$+$30400.0}
\expandafter\def\csname val@rev.desc.d4.uiv.all.downscale-factor.q25\endcsname{0.533}
\expandafter\def\csname cnt@rev.desc.d4.uiv.all.downscale-factor.q25\endcsname{1}
\expandafter\def\csname pct@rev.desc.d4.uiv.all.downscale-factor.q25\endcsname{53.3}
\expandafter\def\csname ppt@rev.desc.d4.uiv.all.downscale-factor.q25\endcsname{$+$53.3}
\expandafter\def\csname val@rev.desc.d4.uiv.all.downscale-factor.q75\endcsname{0.610}
\expandafter\def\csname cnt@rev.desc.d4.uiv.all.downscale-factor.q75\endcsname{1}
\expandafter\def\csname pct@rev.desc.d4.uiv.all.downscale-factor.q75\endcsname{61.0}
\expandafter\def\csname ppt@rev.desc.d4.uiv.all.downscale-factor.q75\endcsname{$+$61.0}
\expandafter\def\csname val@rev.desc.d4.uiv.all.long-edge.median\endcsname{1920.000}
\expandafter\def\csname cnt@rev.desc.d4.uiv.all.long-edge.median\endcsname{1{,}920}
\expandafter\def\csname pct@rev.desc.d4.uiv.all.long-edge.median\endcsname{192000.0}
\expandafter\def\csname ppt@rev.desc.d4.uiv.all.long-edge.median\endcsname{$+$192000.0}
\expandafter\def\csname val@rev.desc.d4.uiv.all.long-edge.n\endcsname{304}
\expandafter\def\csname cnt@rev.desc.d4.uiv.all.long-edge.n\endcsname{304}
\expandafter\def\csname pct@rev.desc.d4.uiv.all.long-edge.n\endcsname{30400.0}
\expandafter\def\csname ppt@rev.desc.d4.uiv.all.long-edge.n\endcsname{$+$30400.0}
\expandafter\def\csname val@rev.desc.d4.uiv.all.long-edge.q25\endcsname{1680.000}
\expandafter\def\csname cnt@rev.desc.d4.uiv.all.long-edge.q25\endcsname{1{,}680}
\expandafter\def\csname pct@rev.desc.d4.uiv.all.long-edge.q25\endcsname{168000.0}
\expandafter\def\csname ppt@rev.desc.d4.uiv.all.long-edge.q25\endcsname{$+$168000.0}
\expandafter\def\csname val@rev.desc.d4.uiv.all.long-edge.q75\endcsname{1920.000}
\expandafter\def\csname cnt@rev.desc.d4.uiv.all.long-edge.q75\endcsname{1{,}920}
\expandafter\def\csname pct@rev.desc.d4.uiv.all.long-edge.q75\endcsname{192000.0}
\expandafter\def\csname ppt@rev.desc.d4.uiv.all.long-edge.q75\endcsname{$+$192000.0}
\expandafter\def\csname val@rev.desc.d4.uiv.all.megapixels.median\endcsname{1.974}
\expandafter\def\csname cnt@rev.desc.d4.uiv.all.megapixels.median\endcsname{2}
\expandafter\def\csname pct@rev.desc.d4.uiv.all.megapixels.median\endcsname{197.4}
\expandafter\def\csname ppt@rev.desc.d4.uiv.all.megapixels.median\endcsname{$+$197.4}
\expandafter\def\csname val@rev.desc.d4.uiv.all.megapixels.n\endcsname{304}
\expandafter\def\csname cnt@rev.desc.d4.uiv.all.megapixels.n\endcsname{304}
\expandafter\def\csname pct@rev.desc.d4.uiv.all.megapixels.n\endcsname{30400.0}
\expandafter\def\csname ppt@rev.desc.d4.uiv.all.megapixels.n\endcsname{$+$30400.0}
\expandafter\def\csname val@rev.desc.d4.uiv.all.megapixels.q25\endcsname{1.754}
\expandafter\def\csname cnt@rev.desc.d4.uiv.all.megapixels.q25\endcsname{2}
\expandafter\def\csname pct@rev.desc.d4.uiv.all.megapixels.q25\endcsname{175.4}
\expandafter\def\csname ppt@rev.desc.d4.uiv.all.megapixels.q25\endcsname{$+$175.4}
\expandafter\def\csname val@rev.desc.d4.uiv.all.megapixels.q75\endcsname{2.074}
\expandafter\def\csname cnt@rev.desc.d4.uiv.all.megapixels.q75\endcsname{2}
\expandafter\def\csname pct@rev.desc.d4.uiv.all.megapixels.q75\endcsname{207.4}
\expandafter\def\csname ppt@rev.desc.d4.uiv.all.megapixels.q75\endcsname{$+$207.4}
\expandafter\def\csname val@rev.desc.d4.uiv.all.n-apps\endcsname{44}
\expandafter\def\csname cnt@rev.desc.d4.uiv.all.n-apps\endcsname{44}
\expandafter\def\csname pct@rev.desc.d4.uiv.all.n-apps\endcsname{4400.0}
\expandafter\def\csname ppt@rev.desc.d4.uiv.all.n-apps\endcsname{$+$4400.0}
\expandafter\def\csname val@rev.desc.d4.uiv.all.n-tasks\endcsname{304}
\expandafter\def\csname cnt@rev.desc.d4.uiv.all.n-tasks\endcsname{304}
\expandafter\def\csname pct@rev.desc.d4.uiv.all.n-tasks\endcsname{30400.0}
\expandafter\def\csname ppt@rev.desc.d4.uiv.all.n-tasks\endcsname{$+$30400.0}
\expandafter\def\csname val@rev.desc.d4.uiv.all.share-downscaled\endcsname{0.997}
\expandafter\def\csname cnt@rev.desc.d4.uiv.all.share-downscaled\endcsname{1}
\expandafter\def\csname pct@rev.desc.d4.uiv.all.share-downscaled\endcsname{99.7}
\expandafter\def\csname ppt@rev.desc.d4.uiv.all.share-downscaled\endcsname{$+$99.7}
\expandafter\def\csname val@rev.desc.d4.uiv.all.share-threshold-gt-target-short-side\endcsname{0.898}
\expandafter\def\csname cnt@rev.desc.d4.uiv.all.share-threshold-gt-target-short-side\endcsname{1}
\expandafter\def\csname pct@rev.desc.d4.uiv.all.share-threshold-gt-target-short-side\endcsname{89.8}
\expandafter\def\csname ppt@rev.desc.d4.uiv.all.share-threshold-gt-target-short-side\endcsname{$+$89.8}
\expandafter\def\csname val@rev.desc.d4.uiv.all.short-edge.median\endcsname{1044.000}
\expandafter\def\csname cnt@rev.desc.d4.uiv.all.short-edge.median\endcsname{1{,}044}
\expandafter\def\csname pct@rev.desc.d4.uiv.all.short-edge.median\endcsname{104400.0}
\expandafter\def\csname ppt@rev.desc.d4.uiv.all.short-edge.median\endcsname{$+$104400.0}
\expandafter\def\csname val@rev.desc.d4.uiv.all.short-edge.n\endcsname{304}
\expandafter\def\csname cnt@rev.desc.d4.uiv.all.short-edge.n\endcsname{304}
\expandafter\def\csname pct@rev.desc.d4.uiv.all.short-edge.n\endcsname{30400.0}
\expandafter\def\csname ppt@rev.desc.d4.uiv.all.short-edge.n\endcsname{$+$30400.0}
\expandafter\def\csname val@rev.desc.d4.uiv.all.short-edge.q25\endcsname{1002.000}
\expandafter\def\csname cnt@rev.desc.d4.uiv.all.short-edge.q25\endcsname{1{,}002}
\expandafter\def\csname pct@rev.desc.d4.uiv.all.short-edge.q25\endcsname{100200.0}
\expandafter\def\csname ppt@rev.desc.d4.uiv.all.short-edge.q25\endcsname{$+$100200.0}
\expandafter\def\csname val@rev.desc.d4.uiv.all.short-edge.q75\endcsname{1080.000}
\expandafter\def\csname cnt@rev.desc.d4.uiv.all.short-edge.q75\endcsname{1{,}080}
\expandafter\def\csname pct@rev.desc.d4.uiv.all.short-edge.q75\endcsname{108000.0}
\expandafter\def\csname ppt@rev.desc.d4.uiv.all.short-edge.q75\endcsname{$+$108000.0}
\expandafter\def\csname val@rev.desc.d4.uiv.all.target-area-fraction.median\endcsname{0.001}
\expandafter\def\csname cnt@rev.desc.d4.uiv.all.target-area-fraction.median\endcsname{0}
\expandafter\def\csname pct@rev.desc.d4.uiv.all.target-area-fraction.median\endcsname{0.1}
\expandafter\def\csname ppt@rev.desc.d4.uiv.all.target-area-fraction.median\endcsname{$+$0.1}
\expandafter\def\csname val@rev.desc.d4.uiv.all.target-area-fraction.n\endcsname{304}
\expandafter\def\csname cnt@rev.desc.d4.uiv.all.target-area-fraction.n\endcsname{304}
\expandafter\def\csname pct@rev.desc.d4.uiv.all.target-area-fraction.n\endcsname{30400.0}
\expandafter\def\csname ppt@rev.desc.d4.uiv.all.target-area-fraction.n\endcsname{$+$30400.0}
\expandafter\def\csname val@rev.desc.d4.uiv.all.target-area-fraction.q25\endcsname{0.000}
\expandafter\def\csname cnt@rev.desc.d4.uiv.all.target-area-fraction.q25\endcsname{0}
\expandafter\def\csname pct@rev.desc.d4.uiv.all.target-area-fraction.q25\endcsname{0.0}
\expandafter\def\csname ppt@rev.desc.d4.uiv.all.target-area-fraction.q25\endcsname{0.0}
\expandafter\def\csname val@rev.desc.d4.uiv.all.target-area-fraction.q75\endcsname{0.001}
\expandafter\def\csname cnt@rev.desc.d4.uiv.all.target-area-fraction.q75\endcsname{0}
\expandafter\def\csname pct@rev.desc.d4.uiv.all.target-area-fraction.q75\endcsname{0.1}
\expandafter\def\csname ppt@rev.desc.d4.uiv.all.target-area-fraction.q75\endcsname{$+$0.1}
\expandafter\def\csname val@rev.desc.d4.uiv.all.target-short-side-direct-view.median\endcsname{15.650}
\expandafter\def\csname cnt@rev.desc.d4.uiv.all.target-short-side-direct-view.median\endcsname{16}
\expandafter\def\csname pct@rev.desc.d4.uiv.all.target-short-side-direct-view.median\endcsname{1565.0}
\expandafter\def\csname ppt@rev.desc.d4.uiv.all.target-short-side-direct-view.median\endcsname{$+$1565.0}
\expandafter\def\csname val@rev.desc.d4.uiv.all.target-short-side-direct-view.n\endcsname{304}
\expandafter\def\csname cnt@rev.desc.d4.uiv.all.target-short-side-direct-view.n\endcsname{304}
\expandafter\def\csname pct@rev.desc.d4.uiv.all.target-short-side-direct-view.n\endcsname{30400.0}
\expandafter\def\csname ppt@rev.desc.d4.uiv.all.target-short-side-direct-view.n\endcsname{$+$30400.0}
\expandafter\def\csname val@rev.desc.d4.uiv.all.target-short-side-direct-view.q25\endcsname{13.143}
\expandafter\def\csname cnt@rev.desc.d4.uiv.all.target-short-side-direct-view.q25\endcsname{13}
\expandafter\def\csname pct@rev.desc.d4.uiv.all.target-short-side-direct-view.q25\endcsname{1314.3}
\expandafter\def\csname ppt@rev.desc.d4.uiv.all.target-short-side-direct-view.q25\endcsname{$+$1314.3}
\expandafter\def\csname val@rev.desc.d4.uiv.all.target-short-side-direct-view.q75\endcsname{19.375}
\expandafter\def\csname cnt@rev.desc.d4.uiv.all.target-short-side-direct-view.q75\endcsname{19}
\expandafter\def\csname pct@rev.desc.d4.uiv.all.target-short-side-direct-view.q75\endcsname{1937.5}
\expandafter\def\csname ppt@rev.desc.d4.uiv.all.target-short-side-direct-view.q75\endcsname{$+$1937.5}
\expandafter\def\csname val@rev.desc.d4.uiv.all.target-short-side-native.median\endcsname{27.771}
\expandafter\def\csname cnt@rev.desc.d4.uiv.all.target-short-side-native.median\endcsname{28}
\expandafter\def\csname pct@rev.desc.d4.uiv.all.target-short-side-native.median\endcsname{2777.1}
\expandafter\def\csname ppt@rev.desc.d4.uiv.all.target-short-side-native.median\endcsname{$+$2777.1}
\expandafter\def\csname val@rev.desc.d4.uiv.all.target-short-side-native.n\endcsname{304}
\expandafter\def\csname cnt@rev.desc.d4.uiv.all.target-short-side-native.n\endcsname{304}
\expandafter\def\csname pct@rev.desc.d4.uiv.all.target-short-side-native.n\endcsname{30400.0}
\expandafter\def\csname ppt@rev.desc.d4.uiv.all.target-short-side-native.n\endcsname{$+$30400.0}
\expandafter\def\csname val@rev.desc.d4.uiv.all.target-short-side-native.q25\endcsname{23.349}
\expandafter\def\csname cnt@rev.desc.d4.uiv.all.target-short-side-native.q25\endcsname{23}
\expandafter\def\csname pct@rev.desc.d4.uiv.all.target-short-side-native.q25\endcsname{2334.9}
\expandafter\def\csname ppt@rev.desc.d4.uiv.all.target-short-side-native.q25\endcsname{$+$2334.9}
\expandafter\def\csname val@rev.desc.d4.uiv.all.target-short-side-native.q75\endcsname{37.029}
\expandafter\def\csname cnt@rev.desc.d4.uiv.all.target-short-side-native.q75\endcsname{37}
\expandafter\def\csname pct@rev.desc.d4.uiv.all.target-short-side-native.q75\endcsname{3702.9}
\expandafter\def\csname ppt@rev.desc.d4.uiv.all.target-short-side-native.q75\endcsname{$+$3702.9}
\expandafter\def\csname val@rev.desc.d4.uiv.census-median-target-side\endcsname{15.650}
\expandafter\def\csname cnt@rev.desc.d4.uiv.census-median-target-side\endcsname{16}
\expandafter\def\csname pct@rev.desc.d4.uiv.census-median-target-side\endcsname{1565.0}
\expandafter\def\csname ppt@rev.desc.d4.uiv.census-median-target-side\endcsname{$+$1565.0}
\expandafter\def\csname val@rev.desc.d4.uiv.reproduces-census-median-target-side\endcsname{1}
\expandafter\def\csname cnt@rev.desc.d4.uiv.reproduces-census-median-target-side\endcsname{1}
\expandafter\def\csname pct@rev.desc.d4.uiv.reproduces-census-median-target-side\endcsname{100.0}
\expandafter\def\csname ppt@rev.desc.d4.uiv.reproduces-census-median-target-side\endcsname{$+$100.0}
\expandafter\def\csname val@rev.desc.d5.annotation-rows.basic\endcsname{1772}
\expandafter\def\csname cnt@rev.desc.d5.annotation-rows.basic\endcsname{1{,}772}
\expandafter\def\csname pct@rev.desc.d5.annotation-rows.basic\endcsname{177200.0}
\expandafter\def\csname ppt@rev.desc.d5.annotation-rows.basic\endcsname{$+$177200.0}
\expandafter\def\csname val@rev.desc.d5.annotation-rows.functional\endcsname{1772}
\expandafter\def\csname cnt@rev.desc.d5.annotation-rows.functional\endcsname{1{,}772}
\expandafter\def\csname pct@rev.desc.d5.annotation-rows.functional\endcsname{177200.0}
\expandafter\def\csname ppt@rev.desc.d5.annotation-rows.functional\endcsname{$+$177200.0}
\expandafter\def\csname val@rev.desc.d5.annotation-rows.spatial-excluded\endcsname{1935}
\expandafter\def\csname cnt@rev.desc.d5.annotation-rows.spatial-excluded\endcsname{1{,}935}
\expandafter\def\csname pct@rev.desc.d5.annotation-rows.spatial-excluded\endcsname{193500.0}
\expandafter\def\csname ppt@rev.desc.d5.annotation-rows.spatial-excluded\endcsname{$+$193500.0}
\expandafter\def\csname val@rev.desc.d5.dropped-invalid-bbox-rows\endcsname{56}
\expandafter\def\csname cnt@rev.desc.d5.dropped-invalid-bbox-rows\endcsname{56}
\expandafter\def\csname pct@rev.desc.d5.dropped-invalid-bbox-rows\endcsname{5600.0}
\expandafter\def\csname ppt@rev.desc.d5.dropped-invalid-bbox-rows\endcsname{$+$5600.0}
\expandafter\def\csname val@rev.desc.d5.groups.kept-1561.element-type-n.icon\endcsname{1215}
\expandafter\def\csname cnt@rev.desc.d5.groups.kept-1561.element-type-n.icon\endcsname{1{,}215}
\expandafter\def\csname pct@rev.desc.d5.groups.kept-1561.element-type-n.icon\endcsname{121500.0}
\expandafter\def\csname ppt@rev.desc.d5.groups.kept-1561.element-type-n.icon\endcsname{$+$121500.0}
\expandafter\def\csname val@rev.desc.d5.groups.kept-1561.element-type-n.text\endcsname{346}
\expandafter\def\csname cnt@rev.desc.d5.groups.kept-1561.element-type-n.text\endcsname{346}
\expandafter\def\csname pct@rev.desc.d5.groups.kept-1561.element-type-n.text\endcsname{34600.0}
\expandafter\def\csname ppt@rev.desc.d5.groups.kept-1561.element-type-n.text\endcsname{$+$34600.0}
\expandafter\def\csname val@rev.desc.d5.groups.kept-1561.element-type-share.icon\endcsname{0.778}
\expandafter\def\csname cnt@rev.desc.d5.groups.kept-1561.element-type-share.icon\endcsname{1}
\expandafter\def\csname pct@rev.desc.d5.groups.kept-1561.element-type-share.icon\endcsname{77.8}
\expandafter\def\csname ppt@rev.desc.d5.groups.kept-1561.element-type-share.icon\endcsname{$+$77.8}
\expandafter\def\csname val@rev.desc.d5.groups.kept-1561.element-type-share.text\endcsname{0.222}
\expandafter\def\csname cnt@rev.desc.d5.groups.kept-1561.element-type-share.text\endcsname{0}
\expandafter\def\csname pct@rev.desc.d5.groups.kept-1561.element-type-share.text\endcsname{22.2}
\expandafter\def\csname ppt@rev.desc.d5.groups.kept-1561.element-type-share.text\endcsname{$+$22.2}
\expandafter\def\csname val@rev.desc.d5.groups.kept-1561.image-long-edge.median\endcsname{1920.000}
\expandafter\def\csname cnt@rev.desc.d5.groups.kept-1561.image-long-edge.median\endcsname{1{,}920}
\expandafter\def\csname pct@rev.desc.d5.groups.kept-1561.image-long-edge.median\endcsname{192000.0}
\expandafter\def\csname ppt@rev.desc.d5.groups.kept-1561.image-long-edge.median\endcsname{$+$192000.0}
\expandafter\def\csname val@rev.desc.d5.groups.kept-1561.image-long-edge.n\endcsname{1561}
\expandafter\def\csname cnt@rev.desc.d5.groups.kept-1561.image-long-edge.n\endcsname{1{,}561}
\expandafter\def\csname pct@rev.desc.d5.groups.kept-1561.image-long-edge.n\endcsname{156100.0}
\expandafter\def\csname ppt@rev.desc.d5.groups.kept-1561.image-long-edge.n\endcsname{$+$156100.0}
\expandafter\def\csname val@rev.desc.d5.groups.kept-1561.image-long-edge.q25\endcsname{1680.000}
\expandafter\def\csname cnt@rev.desc.d5.groups.kept-1561.image-long-edge.q25\endcsname{1{,}680}
\expandafter\def\csname pct@rev.desc.d5.groups.kept-1561.image-long-edge.q25\endcsname{168000.0}
\expandafter\def\csname ppt@rev.desc.d5.groups.kept-1561.image-long-edge.q25\endcsname{$+$168000.0}
\expandafter\def\csname val@rev.desc.d5.groups.kept-1561.image-long-edge.q75\endcsname{1920.000}
\expandafter\def\csname cnt@rev.desc.d5.groups.kept-1561.image-long-edge.q75\endcsname{1{,}920}
\expandafter\def\csname pct@rev.desc.d5.groups.kept-1561.image-long-edge.q75\endcsname{192000.0}
\expandafter\def\csname ppt@rev.desc.d5.groups.kept-1561.image-long-edge.q75\endcsname{$+$192000.0}
\expandafter\def\csname val@rev.desc.d5.groups.kept-1561.n\endcsname{1561}
\expandafter\def\csname cnt@rev.desc.d5.groups.kept-1561.n\endcsname{1{,}561}
\expandafter\def\csname pct@rev.desc.d5.groups.kept-1561.n\endcsname{156100.0}
\expandafter\def\csname ppt@rev.desc.d5.groups.kept-1561.n\endcsname{$+$156100.0}
\expandafter\def\csname val@rev.desc.d5.groups.kept-1561.n-apps\endcsname{83}
\expandafter\def\csname cnt@rev.desc.d5.groups.kept-1561.n-apps\endcsname{83}
\expandafter\def\csname pct@rev.desc.d5.groups.kept-1561.n-apps\endcsname{8300.0}
\expandafter\def\csname ppt@rev.desc.d5.groups.kept-1561.n-apps\endcsname{$+$8300.0}
\expandafter\def\csname val@rev.desc.d5.groups.kept-1561.n-screenshots\endcsname{1164}
\expandafter\def\csname cnt@rev.desc.d5.groups.kept-1561.n-screenshots\endcsname{1{,}164}
\expandafter\def\csname pct@rev.desc.d5.groups.kept-1561.n-screenshots\endcsname{116400.0}
\expandafter\def\csname ppt@rev.desc.d5.groups.kept-1561.n-screenshots\endcsname{$+$116400.0}
\expandafter\def\csname val@rev.desc.d5.groups.kept-1561.share-from-screenshots-with-gt2-roots\endcsname{0.165}
\expandafter\def\csname cnt@rev.desc.d5.groups.kept-1561.share-from-screenshots-with-gt2-roots\endcsname{0}
\expandafter\def\csname pct@rev.desc.d5.groups.kept-1561.share-from-screenshots-with-gt2-roots\endcsname{16.5}
\expandafter\def\csname ppt@rev.desc.d5.groups.kept-1561.share-from-screenshots-with-gt2-roots\endcsname{$+$16.5}
\expandafter\def\csname val@rev.desc.d5.groups.kept-1561.share-with-functional-prompt\endcsname{1.000}
\expandafter\def\csname cnt@rev.desc.d5.groups.kept-1561.share-with-functional-prompt\endcsname{1}
\expandafter\def\csname pct@rev.desc.d5.groups.kept-1561.share-with-functional-prompt\endcsname{100.0}
\expandafter\def\csname ppt@rev.desc.d5.groups.kept-1561.share-with-functional-prompt\endcsname{$+$100.0}
\expandafter\def\csname val@rev.desc.d5.groups.kept-1561.target-area-fraction.median\endcsname{0.001}
\expandafter\def\csname cnt@rev.desc.d5.groups.kept-1561.target-area-fraction.median\endcsname{0}
\expandafter\def\csname pct@rev.desc.d5.groups.kept-1561.target-area-fraction.median\endcsname{0.1}
\expandafter\def\csname ppt@rev.desc.d5.groups.kept-1561.target-area-fraction.median\endcsname{$+$0.1}
\expandafter\def\csname val@rev.desc.d5.groups.kept-1561.target-area-fraction.n\endcsname{1561}
\expandafter\def\csname cnt@rev.desc.d5.groups.kept-1561.target-area-fraction.n\endcsname{1{,}561}
\expandafter\def\csname pct@rev.desc.d5.groups.kept-1561.target-area-fraction.n\endcsname{156100.0}
\expandafter\def\csname ppt@rev.desc.d5.groups.kept-1561.target-area-fraction.n\endcsname{$+$156100.0}
\expandafter\def\csname val@rev.desc.d5.groups.kept-1561.target-area-fraction.q25\endcsname{0.000}
\expandafter\def\csname cnt@rev.desc.d5.groups.kept-1561.target-area-fraction.q25\endcsname{0}
\expandafter\def\csname pct@rev.desc.d5.groups.kept-1561.target-area-fraction.q25\endcsname{0.0}
\expandafter\def\csname ppt@rev.desc.d5.groups.kept-1561.target-area-fraction.q25\endcsname{0.0}
\expandafter\def\csname val@rev.desc.d5.groups.kept-1561.target-area-fraction.q75\endcsname{0.001}
\expandafter\def\csname cnt@rev.desc.d5.groups.kept-1561.target-area-fraction.q75\endcsname{0}
\expandafter\def\csname pct@rev.desc.d5.groups.kept-1561.target-area-fraction.q75\endcsname{0.1}
\expandafter\def\csname ppt@rev.desc.d5.groups.kept-1561.target-area-fraction.q75\endcsname{$+$0.1}
\expandafter\def\csname val@rev.desc.d5.groups.kept-1561.target-short-side-native.median\endcsname{27.300}
\expandafter\def\csname cnt@rev.desc.d5.groups.kept-1561.target-short-side-native.median\endcsname{27}
\expandafter\def\csname pct@rev.desc.d5.groups.kept-1561.target-short-side-native.median\endcsname{2730.0}
\expandafter\def\csname ppt@rev.desc.d5.groups.kept-1561.target-short-side-native.median\endcsname{$+$2730.0}
\expandafter\def\csname val@rev.desc.d5.groups.kept-1561.target-short-side-native.n\endcsname{1561}
\expandafter\def\csname cnt@rev.desc.d5.groups.kept-1561.target-short-side-native.n\endcsname{1{,}561}
\expandafter\def\csname pct@rev.desc.d5.groups.kept-1561.target-short-side-native.n\endcsname{156100.0}
\expandafter\def\csname ppt@rev.desc.d5.groups.kept-1561.target-short-side-native.n\endcsname{$+$156100.0}
\expandafter\def\csname val@rev.desc.d5.groups.kept-1561.target-short-side-native.q25\endcsname{22.305}
\expandafter\def\csname cnt@rev.desc.d5.groups.kept-1561.target-short-side-native.q25\endcsname{22}
\expandafter\def\csname pct@rev.desc.d5.groups.kept-1561.target-short-side-native.q25\endcsname{2230.5}
\expandafter\def\csname ppt@rev.desc.d5.groups.kept-1561.target-short-side-native.q25\endcsname{$+$2230.5}
\expandafter\def\csname val@rev.desc.d5.groups.kept-1561.target-short-side-native.q75\endcsname{34.560}
\expandafter\def\csname cnt@rev.desc.d5.groups.kept-1561.target-short-side-native.q75\endcsname{35}
\expandafter\def\csname pct@rev.desc.d5.groups.kept-1561.target-short-side-native.q75\endcsname{3456.0}
\expandafter\def\csname ppt@rev.desc.d5.groups.kept-1561.target-short-side-native.q75\endcsname{$+$3456.0}
\expandafter\def\csname val@rev.desc.d5.groups.kept-1561.ui-category-share.Button\endcsname{0.541}
\expandafter\def\csname cnt@rev.desc.d5.groups.kept-1561.ui-category-share.Button\endcsname{1}
\expandafter\def\csname pct@rev.desc.d5.groups.kept-1561.ui-category-share.Button\endcsname{54.1}
\expandafter\def\csname ppt@rev.desc.d5.groups.kept-1561.ui-category-share.Button\endcsname{$+$54.1}
\expandafter\def\csname val@rev.desc.d5.groups.kept-1561.ui-category-share.Input Elements\endcsname{0.324}
\expandafter\def\csname cnt@rev.desc.d5.groups.kept-1561.ui-category-share.Input Elements\endcsname{0}
\expandafter\def\csname pct@rev.desc.d5.groups.kept-1561.ui-category-share.Input Elements\endcsname{32.4}
\expandafter\def\csname ppt@rev.desc.d5.groups.kept-1561.ui-category-share.Input Elements\endcsname{$+$32.4}
\expandafter\def\csname val@rev.desc.d5.groups.kept-1561.ui-category-share.Menu\endcsname{0.135}
\expandafter\def\csname cnt@rev.desc.d5.groups.kept-1561.ui-category-share.Menu\endcsname{0}
\expandafter\def\csname pct@rev.desc.d5.groups.kept-1561.ui-category-share.Menu\endcsname{13.5}
\expandafter\def\csname ppt@rev.desc.d5.groups.kept-1561.ui-category-share.Menu\endcsname{$+$13.5}
\expandafter\def\csname val@rev.desc.d5.groups.pool-all.element-type-n.icon\endcsname{1374}
\expandafter\def\csname cnt@rev.desc.d5.groups.pool-all.element-type-n.icon\endcsname{1{,}374}
\expandafter\def\csname pct@rev.desc.d5.groups.pool-all.element-type-n.icon\endcsname{137400.0}
\expandafter\def\csname ppt@rev.desc.d5.groups.pool-all.element-type-n.icon\endcsname{$+$137400.0}
\expandafter\def\csname val@rev.desc.d5.groups.pool-all.element-type-n.text\endcsname{370}
\expandafter\def\csname cnt@rev.desc.d5.groups.pool-all.element-type-n.text\endcsname{370}
\expandafter\def\csname pct@rev.desc.d5.groups.pool-all.element-type-n.text\endcsname{37000.0}
\expandafter\def\csname ppt@rev.desc.d5.groups.pool-all.element-type-n.text\endcsname{$+$37000.0}
\expandafter\def\csname val@rev.desc.d5.groups.pool-all.element-type-share.icon\endcsname{0.788}
\expandafter\def\csname cnt@rev.desc.d5.groups.pool-all.element-type-share.icon\endcsname{1}
\expandafter\def\csname pct@rev.desc.d5.groups.pool-all.element-type-share.icon\endcsname{78.8}
\expandafter\def\csname ppt@rev.desc.d5.groups.pool-all.element-type-share.icon\endcsname{$+$78.8}
\expandafter\def\csname val@rev.desc.d5.groups.pool-all.element-type-share.text\endcsname{0.212}
\expandafter\def\csname cnt@rev.desc.d5.groups.pool-all.element-type-share.text\endcsname{0}
\expandafter\def\csname pct@rev.desc.d5.groups.pool-all.element-type-share.text\endcsname{21.2}
\expandafter\def\csname ppt@rev.desc.d5.groups.pool-all.element-type-share.text\endcsname{$+$21.2}
\expandafter\def\csname val@rev.desc.d5.groups.pool-all.image-long-edge.median\endcsname{1920.000}
\expandafter\def\csname cnt@rev.desc.d5.groups.pool-all.image-long-edge.median\endcsname{1{,}920}
\expandafter\def\csname pct@rev.desc.d5.groups.pool-all.image-long-edge.median\endcsname{192000.0}
\expandafter\def\csname ppt@rev.desc.d5.groups.pool-all.image-long-edge.median\endcsname{$+$192000.0}
\expandafter\def\csname val@rev.desc.d5.groups.pool-all.image-long-edge.n\endcsname{1744}
\expandafter\def\csname cnt@rev.desc.d5.groups.pool-all.image-long-edge.n\endcsname{1{,}744}
\expandafter\def\csname pct@rev.desc.d5.groups.pool-all.image-long-edge.n\endcsname{174400.0}
\expandafter\def\csname ppt@rev.desc.d5.groups.pool-all.image-long-edge.n\endcsname{$+$174400.0}
\expandafter\def\csname val@rev.desc.d5.groups.pool-all.image-long-edge.q25\endcsname{1680.000}
\expandafter\def\csname cnt@rev.desc.d5.groups.pool-all.image-long-edge.q25\endcsname{1{,}680}
\expandafter\def\csname pct@rev.desc.d5.groups.pool-all.image-long-edge.q25\endcsname{168000.0}
\expandafter\def\csname ppt@rev.desc.d5.groups.pool-all.image-long-edge.q25\endcsname{$+$168000.0}
\expandafter\def\csname val@rev.desc.d5.groups.pool-all.image-long-edge.q75\endcsname{1920.000}
\expandafter\def\csname cnt@rev.desc.d5.groups.pool-all.image-long-edge.q75\endcsname{1{,}920}
\expandafter\def\csname pct@rev.desc.d5.groups.pool-all.image-long-edge.q75\endcsname{192000.0}
\expandafter\def\csname ppt@rev.desc.d5.groups.pool-all.image-long-edge.q75\endcsname{$+$192000.0}
\expandafter\def\csname val@rev.desc.d5.groups.pool-all.n\endcsname{1744}
\expandafter\def\csname cnt@rev.desc.d5.groups.pool-all.n\endcsname{1{,}744}
\expandafter\def\csname pct@rev.desc.d5.groups.pool-all.n\endcsname{174400.0}
\expandafter\def\csname ppt@rev.desc.d5.groups.pool-all.n\endcsname{$+$174400.0}
\expandafter\def\csname val@rev.desc.d5.groups.pool-all.n-apps\endcsname{83}
\expandafter\def\csname cnt@rev.desc.d5.groups.pool-all.n-apps\endcsname{83}
\expandafter\def\csname pct@rev.desc.d5.groups.pool-all.n-apps\endcsname{8300.0}
\expandafter\def\csname ppt@rev.desc.d5.groups.pool-all.n-apps\endcsname{$+$8300.0}
\expandafter\def\csname val@rev.desc.d5.groups.pool-all.n-screenshots\endcsname{1164}
\expandafter\def\csname cnt@rev.desc.d5.groups.pool-all.n-screenshots\endcsname{1{,}164}
\expandafter\def\csname pct@rev.desc.d5.groups.pool-all.n-screenshots\endcsname{116400.0}
\expandafter\def\csname ppt@rev.desc.d5.groups.pool-all.n-screenshots\endcsname{$+$116400.0}
\expandafter\def\csname val@rev.desc.d5.groups.pool-all.share-from-screenshots-with-gt2-roots\endcsname{0.253}
\expandafter\def\csname cnt@rev.desc.d5.groups.pool-all.share-from-screenshots-with-gt2-roots\endcsname{0}
\expandafter\def\csname pct@rev.desc.d5.groups.pool-all.share-from-screenshots-with-gt2-roots\endcsname{25.3}
\expandafter\def\csname ppt@rev.desc.d5.groups.pool-all.share-from-screenshots-with-gt2-roots\endcsname{$+$25.3}
\expandafter\def\csname val@rev.desc.d5.groups.pool-all.share-with-functional-prompt\endcsname{1.000}
\expandafter\def\csname cnt@rev.desc.d5.groups.pool-all.share-with-functional-prompt\endcsname{1}
\expandafter\def\csname pct@rev.desc.d5.groups.pool-all.share-with-functional-prompt\endcsname{100.0}
\expandafter\def\csname ppt@rev.desc.d5.groups.pool-all.share-with-functional-prompt\endcsname{$+$100.0}
\expandafter\def\csname val@rev.desc.d5.groups.pool-all.target-area-fraction.median\endcsname{0.001}
\expandafter\def\csname cnt@rev.desc.d5.groups.pool-all.target-area-fraction.median\endcsname{0}
\expandafter\def\csname pct@rev.desc.d5.groups.pool-all.target-area-fraction.median\endcsname{0.1}
\expandafter\def\csname ppt@rev.desc.d5.groups.pool-all.target-area-fraction.median\endcsname{$+$0.1}
\expandafter\def\csname val@rev.desc.d5.groups.pool-all.target-area-fraction.n\endcsname{1744}
\expandafter\def\csname cnt@rev.desc.d5.groups.pool-all.target-area-fraction.n\endcsname{1{,}744}
\expandafter\def\csname pct@rev.desc.d5.groups.pool-all.target-area-fraction.n\endcsname{174400.0}
\expandafter\def\csname ppt@rev.desc.d5.groups.pool-all.target-area-fraction.n\endcsname{$+$174400.0}
\expandafter\def\csname val@rev.desc.d5.groups.pool-all.target-area-fraction.q25\endcsname{0.000}
\expandafter\def\csname cnt@rev.desc.d5.groups.pool-all.target-area-fraction.q25\endcsname{0}
\expandafter\def\csname pct@rev.desc.d5.groups.pool-all.target-area-fraction.q25\endcsname{0.0}
\expandafter\def\csname ppt@rev.desc.d5.groups.pool-all.target-area-fraction.q25\endcsname{0.0}
\expandafter\def\csname val@rev.desc.d5.groups.pool-all.target-area-fraction.q75\endcsname{0.001}
\expandafter\def\csname cnt@rev.desc.d5.groups.pool-all.target-area-fraction.q75\endcsname{0}
\expandafter\def\csname pct@rev.desc.d5.groups.pool-all.target-area-fraction.q75\endcsname{0.1}
\expandafter\def\csname ppt@rev.desc.d5.groups.pool-all.target-area-fraction.q75\endcsname{$+$0.1}
\expandafter\def\csname val@rev.desc.d5.groups.pool-all.target-short-side-native.median\endcsname{27.174}
\expandafter\def\csname cnt@rev.desc.d5.groups.pool-all.target-short-side-native.median\endcsname{27}
\expandafter\def\csname pct@rev.desc.d5.groups.pool-all.target-short-side-native.median\endcsname{2717.4}
\expandafter\def\csname ppt@rev.desc.d5.groups.pool-all.target-short-side-native.median\endcsname{$+$2717.4}
\expandafter\def\csname val@rev.desc.d5.groups.pool-all.target-short-side-native.n\endcsname{1744}
\expandafter\def\csname cnt@rev.desc.d5.groups.pool-all.target-short-side-native.n\endcsname{1{,}744}
\expandafter\def\csname pct@rev.desc.d5.groups.pool-all.target-short-side-native.n\endcsname{174400.0}
\expandafter\def\csname ppt@rev.desc.d5.groups.pool-all.target-short-side-native.n\endcsname{$+$174400.0}
\expandafter\def\csname val@rev.desc.d5.groups.pool-all.target-short-side-native.q25\endcsname{22.340}
\expandafter\def\csname cnt@rev.desc.d5.groups.pool-all.target-short-side-native.q25\endcsname{22}
\expandafter\def\csname pct@rev.desc.d5.groups.pool-all.target-short-side-native.q25\endcsname{2234.0}
\expandafter\def\csname ppt@rev.desc.d5.groups.pool-all.target-short-side-native.q25\endcsname{$+$2234.0}
\expandafter\def\csname val@rev.desc.d5.groups.pool-all.target-short-side-native.q75\endcsname{34.303}
\expandafter\def\csname cnt@rev.desc.d5.groups.pool-all.target-short-side-native.q75\endcsname{34}
\expandafter\def\csname pct@rev.desc.d5.groups.pool-all.target-short-side-native.q75\endcsname{3430.3}
\expandafter\def\csname ppt@rev.desc.d5.groups.pool-all.target-short-side-native.q75\endcsname{$+$3430.3}
\expandafter\def\csname val@rev.desc.d5.groups.pool-all.ui-category-share.Button\endcsname{0.539}
\expandafter\def\csname cnt@rev.desc.d5.groups.pool-all.ui-category-share.Button\endcsname{1}
\expandafter\def\csname pct@rev.desc.d5.groups.pool-all.ui-category-share.Button\endcsname{53.9}
\expandafter\def\csname ppt@rev.desc.d5.groups.pool-all.ui-category-share.Button\endcsname{$+$53.9}
\expandafter\def\csname val@rev.desc.d5.groups.pool-all.ui-category-share.Input Elements\endcsname{0.324}
\expandafter\def\csname cnt@rev.desc.d5.groups.pool-all.ui-category-share.Input Elements\endcsname{0}
\expandafter\def\csname pct@rev.desc.d5.groups.pool-all.ui-category-share.Input Elements\endcsname{32.4}
\expandafter\def\csname ppt@rev.desc.d5.groups.pool-all.ui-category-share.Input Elements\endcsname{$+$32.4}
\expandafter\def\csname val@rev.desc.d5.groups.pool-all.ui-category-share.Menu\endcsname{0.137}
\expandafter\def\csname cnt@rev.desc.d5.groups.pool-all.ui-category-share.Menu\endcsname{0}
\expandafter\def\csname pct@rev.desc.d5.groups.pool-all.ui-category-share.Menu\endcsname{13.7}
\expandafter\def\csname ppt@rev.desc.d5.groups.pool-all.ui-category-share.Menu\endcsname{$+$13.7}
\expandafter\def\csname val@rev.desc.d5.groups.removed-by-cap-or-prompt.element-type-n.icon\endcsname{159}
\expandafter\def\csname cnt@rev.desc.d5.groups.removed-by-cap-or-prompt.element-type-n.icon\endcsname{159}
\expandafter\def\csname pct@rev.desc.d5.groups.removed-by-cap-or-prompt.element-type-n.icon\endcsname{15900.0}
\expandafter\def\csname ppt@rev.desc.d5.groups.removed-by-cap-or-prompt.element-type-n.icon\endcsname{$+$15900.0}
\expandafter\def\csname val@rev.desc.d5.groups.removed-by-cap-or-prompt.element-type-n.text\endcsname{24}
\expandafter\def\csname cnt@rev.desc.d5.groups.removed-by-cap-or-prompt.element-type-n.text\endcsname{24}
\expandafter\def\csname pct@rev.desc.d5.groups.removed-by-cap-or-prompt.element-type-n.text\endcsname{2400.0}
\expandafter\def\csname ppt@rev.desc.d5.groups.removed-by-cap-or-prompt.element-type-n.text\endcsname{$+$2400.0}
\expandafter\def\csname val@rev.desc.d5.groups.removed-by-cap-or-prompt.element-type-share.icon\endcsname{0.869}
\expandafter\def\csname cnt@rev.desc.d5.groups.removed-by-cap-or-prompt.element-type-share.icon\endcsname{1}
\expandafter\def\csname pct@rev.desc.d5.groups.removed-by-cap-or-prompt.element-type-share.icon\endcsname{86.9}
\expandafter\def\csname ppt@rev.desc.d5.groups.removed-by-cap-or-prompt.element-type-share.icon\endcsname{$+$86.9}
\expandafter\def\csname val@rev.desc.d5.groups.removed-by-cap-or-prompt.element-type-share.text\endcsname{0.131}
\expandafter\def\csname cnt@rev.desc.d5.groups.removed-by-cap-or-prompt.element-type-share.text\endcsname{0}
\expandafter\def\csname pct@rev.desc.d5.groups.removed-by-cap-or-prompt.element-type-share.text\endcsname{13.1}
\expandafter\def\csname ppt@rev.desc.d5.groups.removed-by-cap-or-prompt.element-type-share.text\endcsname{$+$13.1}
\expandafter\def\csname val@rev.desc.d5.groups.removed-by-cap-or-prompt.image-long-edge.median\endcsname{1920.000}
\expandafter\def\csname cnt@rev.desc.d5.groups.removed-by-cap-or-prompt.image-long-edge.median\endcsname{1{,}920}
\expandafter\def\csname pct@rev.desc.d5.groups.removed-by-cap-or-prompt.image-long-edge.median\endcsname{192000.0}
\expandafter\def\csname ppt@rev.desc.d5.groups.removed-by-cap-or-prompt.image-long-edge.median\endcsname{$+$192000.0}
\expandafter\def\csname val@rev.desc.d5.groups.removed-by-cap-or-prompt.image-long-edge.n\endcsname{183}
\expandafter\def\csname cnt@rev.desc.d5.groups.removed-by-cap-or-prompt.image-long-edge.n\endcsname{183}
\expandafter\def\csname pct@rev.desc.d5.groups.removed-by-cap-or-prompt.image-long-edge.n\endcsname{18300.0}
\expandafter\def\csname ppt@rev.desc.d5.groups.removed-by-cap-or-prompt.image-long-edge.n\endcsname{$+$18300.0}
\expandafter\def\csname val@rev.desc.d5.groups.removed-by-cap-or-prompt.image-long-edge.q25\endcsname{1680.000}
\expandafter\def\csname cnt@rev.desc.d5.groups.removed-by-cap-or-prompt.image-long-edge.q25\endcsname{1{,}680}
\expandafter\def\csname pct@rev.desc.d5.groups.removed-by-cap-or-prompt.image-long-edge.q25\endcsname{168000.0}
\expandafter\def\csname ppt@rev.desc.d5.groups.removed-by-cap-or-prompt.image-long-edge.q25\endcsname{$+$168000.0}
\expandafter\def\csname val@rev.desc.d5.groups.removed-by-cap-or-prompt.image-long-edge.q75\endcsname{1920.000}
\expandafter\def\csname cnt@rev.desc.d5.groups.removed-by-cap-or-prompt.image-long-edge.q75\endcsname{1{,}920}
\expandafter\def\csname pct@rev.desc.d5.groups.removed-by-cap-or-prompt.image-long-edge.q75\endcsname{192000.0}
\expandafter\def\csname ppt@rev.desc.d5.groups.removed-by-cap-or-prompt.image-long-edge.q75\endcsname{$+$192000.0}
\expandafter\def\csname val@rev.desc.d5.groups.removed-by-cap-or-prompt.n\endcsname{183}
\expandafter\def\csname cnt@rev.desc.d5.groups.removed-by-cap-or-prompt.n\endcsname{183}
\expandafter\def\csname pct@rev.desc.d5.groups.removed-by-cap-or-prompt.n\endcsname{18300.0}
\expandafter\def\csname ppt@rev.desc.d5.groups.removed-by-cap-or-prompt.n\endcsname{$+$18300.0}
\expandafter\def\csname val@rev.desc.d5.groups.removed-by-cap-or-prompt.n-apps\endcsname{45}
\expandafter\def\csname cnt@rev.desc.d5.groups.removed-by-cap-or-prompt.n-apps\endcsname{45}
\expandafter\def\csname pct@rev.desc.d5.groups.removed-by-cap-or-prompt.n-apps\endcsname{4500.0}
\expandafter\def\csname ppt@rev.desc.d5.groups.removed-by-cap-or-prompt.n-apps\endcsname{$+$4500.0}
\expandafter\def\csname val@rev.desc.d5.groups.removed-by-cap-or-prompt.n-screenshots\endcsname{129}
\expandafter\def\csname cnt@rev.desc.d5.groups.removed-by-cap-or-prompt.n-screenshots\endcsname{129}
\expandafter\def\csname pct@rev.desc.d5.groups.removed-by-cap-or-prompt.n-screenshots\endcsname{12900.0}
\expandafter\def\csname ppt@rev.desc.d5.groups.removed-by-cap-or-prompt.n-screenshots\endcsname{$+$12900.0}
\expandafter\def\csname val@rev.desc.d5.groups.removed-by-cap-or-prompt.share-from-screenshots-with-gt2-roots\endcsname{1.000}
\expandafter\def\csname cnt@rev.desc.d5.groups.removed-by-cap-or-prompt.share-from-screenshots-with-gt2-roots\endcsname{1}
\expandafter\def\csname pct@rev.desc.d5.groups.removed-by-cap-or-prompt.share-from-screenshots-with-gt2-roots\endcsname{100.0}
\expandafter\def\csname ppt@rev.desc.d5.groups.removed-by-cap-or-prompt.share-from-screenshots-with-gt2-roots\endcsname{$+$100.0}
\expandafter\def\csname val@rev.desc.d5.groups.removed-by-cap-or-prompt.share-with-functional-prompt\endcsname{1.000}
\expandafter\def\csname cnt@rev.desc.d5.groups.removed-by-cap-or-prompt.share-with-functional-prompt\endcsname{1}
\expandafter\def\csname pct@rev.desc.d5.groups.removed-by-cap-or-prompt.share-with-functional-prompt\endcsname{100.0}
\expandafter\def\csname ppt@rev.desc.d5.groups.removed-by-cap-or-prompt.share-with-functional-prompt\endcsname{$+$100.0}
\expandafter\def\csname val@rev.desc.d5.groups.removed-by-cap-or-prompt.target-area-fraction.median\endcsname{0.001}
\expandafter\def\csname cnt@rev.desc.d5.groups.removed-by-cap-or-prompt.target-area-fraction.median\endcsname{0}
\expandafter\def\csname pct@rev.desc.d5.groups.removed-by-cap-or-prompt.target-area-fraction.median\endcsname{0.1}
\expandafter\def\csname ppt@rev.desc.d5.groups.removed-by-cap-or-prompt.target-area-fraction.median\endcsname{$+$0.1}
\expandafter\def\csname val@rev.desc.d5.groups.removed-by-cap-or-prompt.target-area-fraction.n\endcsname{183}
\expandafter\def\csname cnt@rev.desc.d5.groups.removed-by-cap-or-prompt.target-area-fraction.n\endcsname{183}
\expandafter\def\csname pct@rev.desc.d5.groups.removed-by-cap-or-prompt.target-area-fraction.n\endcsname{18300.0}
\expandafter\def\csname ppt@rev.desc.d5.groups.removed-by-cap-or-prompt.target-area-fraction.n\endcsname{$+$18300.0}
\expandafter\def\csname val@rev.desc.d5.groups.removed-by-cap-or-prompt.target-area-fraction.q25\endcsname{0.000}
\expandafter\def\csname cnt@rev.desc.d5.groups.removed-by-cap-or-prompt.target-area-fraction.q25\endcsname{0}
\expandafter\def\csname pct@rev.desc.d5.groups.removed-by-cap-or-prompt.target-area-fraction.q25\endcsname{0.0}
\expandafter\def\csname ppt@rev.desc.d5.groups.removed-by-cap-or-prompt.target-area-fraction.q25\endcsname{0.0}
\expandafter\def\csname val@rev.desc.d5.groups.removed-by-cap-or-prompt.target-area-fraction.q75\endcsname{0.001}
\expandafter\def\csname cnt@rev.desc.d5.groups.removed-by-cap-or-prompt.target-area-fraction.q75\endcsname{0}
\expandafter\def\csname pct@rev.desc.d5.groups.removed-by-cap-or-prompt.target-area-fraction.q75\endcsname{0.1}
\expandafter\def\csname ppt@rev.desc.d5.groups.removed-by-cap-or-prompt.target-area-fraction.q75\endcsname{$+$0.1}
\expandafter\def\csname val@rev.desc.d5.groups.removed-by-cap-or-prompt.target-short-side-native.median\endcsname{26.448}
\expandafter\def\csname cnt@rev.desc.d5.groups.removed-by-cap-or-prompt.target-short-side-native.median\endcsname{26}
\expandafter\def\csname pct@rev.desc.d5.groups.removed-by-cap-or-prompt.target-short-side-native.median\endcsname{2644.8}
\expandafter\def\csname ppt@rev.desc.d5.groups.removed-by-cap-or-prompt.target-short-side-native.median\endcsname{$+$2644.8}
\expandafter\def\csname val@rev.desc.d5.groups.removed-by-cap-or-prompt.target-short-side-native.n\endcsname{183}
\expandafter\def\csname cnt@rev.desc.d5.groups.removed-by-cap-or-prompt.target-short-side-native.n\endcsname{183}
\expandafter\def\csname pct@rev.desc.d5.groups.removed-by-cap-or-prompt.target-short-side-native.n\endcsname{18300.0}
\expandafter\def\csname ppt@rev.desc.d5.groups.removed-by-cap-or-prompt.target-short-side-native.n\endcsname{$+$18300.0}
\expandafter\def\csname val@rev.desc.d5.groups.removed-by-cap-or-prompt.target-short-side-native.q25\endcsname{22.371}
\expandafter\def\csname cnt@rev.desc.d5.groups.removed-by-cap-or-prompt.target-short-side-native.q25\endcsname{22}
\expandafter\def\csname pct@rev.desc.d5.groups.removed-by-cap-or-prompt.target-short-side-native.q25\endcsname{2237.1}
\expandafter\def\csname ppt@rev.desc.d5.groups.removed-by-cap-or-prompt.target-short-side-native.q25\endcsname{$+$2237.1}
\expandafter\def\csname val@rev.desc.d5.groups.removed-by-cap-or-prompt.target-short-side-native.q75\endcsname{32.811}
\expandafter\def\csname cnt@rev.desc.d5.groups.removed-by-cap-or-prompt.target-short-side-native.q75\endcsname{33}
\expandafter\def\csname pct@rev.desc.d5.groups.removed-by-cap-or-prompt.target-short-side-native.q75\endcsname{3281.1}
\expandafter\def\csname ppt@rev.desc.d5.groups.removed-by-cap-or-prompt.target-short-side-native.q75\endcsname{$+$3281.1}
\expandafter\def\csname val@rev.desc.d5.groups.removed-by-cap-or-prompt.ui-category-share.Button\endcsname{0.519}
\expandafter\def\csname cnt@rev.desc.d5.groups.removed-by-cap-or-prompt.ui-category-share.Button\endcsname{1}
\expandafter\def\csname pct@rev.desc.d5.groups.removed-by-cap-or-prompt.ui-category-share.Button\endcsname{51.9}
\expandafter\def\csname ppt@rev.desc.d5.groups.removed-by-cap-or-prompt.ui-category-share.Button\endcsname{$+$51.9}
\expandafter\def\csname val@rev.desc.d5.groups.removed-by-cap-or-prompt.ui-category-share.Input Elements\endcsname{0.322}
\expandafter\def\csname cnt@rev.desc.d5.groups.removed-by-cap-or-prompt.ui-category-share.Input Elements\endcsname{0}
\expandafter\def\csname pct@rev.desc.d5.groups.removed-by-cap-or-prompt.ui-category-share.Input Elements\endcsname{32.2}
\expandafter\def\csname ppt@rev.desc.d5.groups.removed-by-cap-or-prompt.ui-category-share.Input Elements\endcsname{$+$32.2}
\expandafter\def\csname val@rev.desc.d5.groups.removed-by-cap-or-prompt.ui-category-share.Menu\endcsname{0.158}
\expandafter\def\csname cnt@rev.desc.d5.groups.removed-by-cap-or-prompt.ui-category-share.Menu\endcsname{0}
\expandafter\def\csname pct@rev.desc.d5.groups.removed-by-cap-or-prompt.ui-category-share.Menu\endcsname{15.8}
\expandafter\def\csname ppt@rev.desc.d5.groups.removed-by-cap-or-prompt.ui-category-share.Menu\endcsname{$+$15.8}
\expandafter\def\csname val@rev.desc.d5.groups.roles-ACT-before-reduction.element-type-n.icon\endcsname{749}
\expandafter\def\csname cnt@rev.desc.d5.groups.roles-ACT-before-reduction.element-type-n.icon\endcsname{749}
\expandafter\def\csname pct@rev.desc.d5.groups.roles-ACT-before-reduction.element-type-n.icon\endcsname{74900.0}
\expandafter\def\csname ppt@rev.desc.d5.groups.roles-ACT-before-reduction.element-type-n.icon\endcsname{$+$74900.0}
\expandafter\def\csname val@rev.desc.d5.groups.roles-ACT-before-reduction.element-type-n.text\endcsname{194}
\expandafter\def\csname cnt@rev.desc.d5.groups.roles-ACT-before-reduction.element-type-n.text\endcsname{194}
\expandafter\def\csname pct@rev.desc.d5.groups.roles-ACT-before-reduction.element-type-n.text\endcsname{19400.0}
\expandafter\def\csname ppt@rev.desc.d5.groups.roles-ACT-before-reduction.element-type-n.text\endcsname{$+$19400.0}
\expandafter\def\csname val@rev.desc.d5.groups.roles-ACT-before-reduction.element-type-share.icon\endcsname{0.794}
\expandafter\def\csname cnt@rev.desc.d5.groups.roles-ACT-before-reduction.element-type-share.icon\endcsname{1}
\expandafter\def\csname pct@rev.desc.d5.groups.roles-ACT-before-reduction.element-type-share.icon\endcsname{79.4}
\expandafter\def\csname ppt@rev.desc.d5.groups.roles-ACT-before-reduction.element-type-share.icon\endcsname{$+$79.4}
\expandafter\def\csname val@rev.desc.d5.groups.roles-ACT-before-reduction.element-type-share.text\endcsname{0.206}
\expandafter\def\csname cnt@rev.desc.d5.groups.roles-ACT-before-reduction.element-type-share.text\endcsname{0}
\expandafter\def\csname pct@rev.desc.d5.groups.roles-ACT-before-reduction.element-type-share.text\endcsname{20.6}
\expandafter\def\csname ppt@rev.desc.d5.groups.roles-ACT-before-reduction.element-type-share.text\endcsname{$+$20.6}
\expandafter\def\csname val@rev.desc.d5.groups.roles-ACT-before-reduction.image-long-edge.median\endcsname{1920.000}
\expandafter\def\csname cnt@rev.desc.d5.groups.roles-ACT-before-reduction.image-long-edge.median\endcsname{1{,}920}
\expandafter\def\csname pct@rev.desc.d5.groups.roles-ACT-before-reduction.image-long-edge.median\endcsname{192000.0}
\expandafter\def\csname ppt@rev.desc.d5.groups.roles-ACT-before-reduction.image-long-edge.median\endcsname{$+$192000.0}
\expandafter\def\csname val@rev.desc.d5.groups.roles-ACT-before-reduction.image-long-edge.n\endcsname{943}
\expandafter\def\csname cnt@rev.desc.d5.groups.roles-ACT-before-reduction.image-long-edge.n\endcsname{943}
\expandafter\def\csname pct@rev.desc.d5.groups.roles-ACT-before-reduction.image-long-edge.n\endcsname{94300.0}
\expandafter\def\csname ppt@rev.desc.d5.groups.roles-ACT-before-reduction.image-long-edge.n\endcsname{$+$94300.0}
\expandafter\def\csname val@rev.desc.d5.groups.roles-ACT-before-reduction.image-long-edge.q25\endcsname{1680.000}
\expandafter\def\csname cnt@rev.desc.d5.groups.roles-ACT-before-reduction.image-long-edge.q25\endcsname{1{,}680}
\expandafter\def\csname pct@rev.desc.d5.groups.roles-ACT-before-reduction.image-long-edge.q25\endcsname{168000.0}
\expandafter\def\csname ppt@rev.desc.d5.groups.roles-ACT-before-reduction.image-long-edge.q25\endcsname{$+$168000.0}
\expandafter\def\csname val@rev.desc.d5.groups.roles-ACT-before-reduction.image-long-edge.q75\endcsname{1920.000}
\expandafter\def\csname cnt@rev.desc.d5.groups.roles-ACT-before-reduction.image-long-edge.q75\endcsname{1{,}920}
\expandafter\def\csname pct@rev.desc.d5.groups.roles-ACT-before-reduction.image-long-edge.q75\endcsname{192000.0}
\expandafter\def\csname ppt@rev.desc.d5.groups.roles-ACT-before-reduction.image-long-edge.q75\endcsname{$+$192000.0}
\expandafter\def\csname val@rev.desc.d5.groups.roles-ACT-before-reduction.n\endcsname{943}
\expandafter\def\csname cnt@rev.desc.d5.groups.roles-ACT-before-reduction.n\endcsname{943}
\expandafter\def\csname pct@rev.desc.d5.groups.roles-ACT-before-reduction.n\endcsname{94300.0}
\expandafter\def\csname ppt@rev.desc.d5.groups.roles-ACT-before-reduction.n\endcsname{$+$94300.0}
\expandafter\def\csname val@rev.desc.d5.groups.roles-ACT-before-reduction.n-apps\endcsname{44}
\expandafter\def\csname cnt@rev.desc.d5.groups.roles-ACT-before-reduction.n-apps\endcsname{44}
\expandafter\def\csname pct@rev.desc.d5.groups.roles-ACT-before-reduction.n-apps\endcsname{4400.0}
\expandafter\def\csname ppt@rev.desc.d5.groups.roles-ACT-before-reduction.n-apps\endcsname{$+$4400.0}
\expandafter\def\csname val@rev.desc.d5.groups.roles-ACT-before-reduction.n-screenshots\endcsname{701}
\expandafter\def\csname cnt@rev.desc.d5.groups.roles-ACT-before-reduction.n-screenshots\endcsname{701}
\expandafter\def\csname pct@rev.desc.d5.groups.roles-ACT-before-reduction.n-screenshots\endcsname{70100.0}
\expandafter\def\csname ppt@rev.desc.d5.groups.roles-ACT-before-reduction.n-screenshots\endcsname{$+$70100.0}
\expandafter\def\csname val@rev.desc.d5.groups.roles-ACT-before-reduction.share-from-screenshots-with-gt2-roots\endcsname{0.157}
\expandafter\def\csname cnt@rev.desc.d5.groups.roles-ACT-before-reduction.share-from-screenshots-with-gt2-roots\endcsname{0}
\expandafter\def\csname pct@rev.desc.d5.groups.roles-ACT-before-reduction.share-from-screenshots-with-gt2-roots\endcsname{15.7}
\expandafter\def\csname ppt@rev.desc.d5.groups.roles-ACT-before-reduction.share-from-screenshots-with-gt2-roots\endcsname{$+$15.7}
\expandafter\def\csname val@rev.desc.d5.groups.roles-ACT-before-reduction.share-with-functional-prompt\endcsname{1.000}
\expandafter\def\csname cnt@rev.desc.d5.groups.roles-ACT-before-reduction.share-with-functional-prompt\endcsname{1}
\expandafter\def\csname pct@rev.desc.d5.groups.roles-ACT-before-reduction.share-with-functional-prompt\endcsname{100.0}
\expandafter\def\csname ppt@rev.desc.d5.groups.roles-ACT-before-reduction.share-with-functional-prompt\endcsname{$+$100.0}
\expandafter\def\csname val@rev.desc.d5.groups.roles-ACT-before-reduction.target-area-fraction.median\endcsname{0.001}
\expandafter\def\csname cnt@rev.desc.d5.groups.roles-ACT-before-reduction.target-area-fraction.median\endcsname{0}
\expandafter\def\csname pct@rev.desc.d5.groups.roles-ACT-before-reduction.target-area-fraction.median\endcsname{0.1}
\expandafter\def\csname ppt@rev.desc.d5.groups.roles-ACT-before-reduction.target-area-fraction.median\endcsname{$+$0.1}
\expandafter\def\csname val@rev.desc.d5.groups.roles-ACT-before-reduction.target-area-fraction.n\endcsname{943}
\expandafter\def\csname cnt@rev.desc.d5.groups.roles-ACT-before-reduction.target-area-fraction.n\endcsname{943}
\expandafter\def\csname pct@rev.desc.d5.groups.roles-ACT-before-reduction.target-area-fraction.n\endcsname{94300.0}
\expandafter\def\csname ppt@rev.desc.d5.groups.roles-ACT-before-reduction.target-area-fraction.n\endcsname{$+$94300.0}
\expandafter\def\csname val@rev.desc.d5.groups.roles-ACT-before-reduction.target-area-fraction.q25\endcsname{0.000}
\expandafter\def\csname cnt@rev.desc.d5.groups.roles-ACT-before-reduction.target-area-fraction.q25\endcsname{0}
\expandafter\def\csname pct@rev.desc.d5.groups.roles-ACT-before-reduction.target-area-fraction.q25\endcsname{0.0}
\expandafter\def\csname ppt@rev.desc.d5.groups.roles-ACT-before-reduction.target-area-fraction.q25\endcsname{0.0}
\expandafter\def\csname val@rev.desc.d5.groups.roles-ACT-before-reduction.target-area-fraction.q75\endcsname{0.001}
\expandafter\def\csname cnt@rev.desc.d5.groups.roles-ACT-before-reduction.target-area-fraction.q75\endcsname{0}
\expandafter\def\csname pct@rev.desc.d5.groups.roles-ACT-before-reduction.target-area-fraction.q75\endcsname{0.1}
\expandafter\def\csname ppt@rev.desc.d5.groups.roles-ACT-before-reduction.target-area-fraction.q75\endcsname{$+$0.1}
\expandafter\def\csname val@rev.desc.d5.groups.roles-ACT-before-reduction.target-short-side-native.median\endcsname{27.200}
\expandafter\def\csname cnt@rev.desc.d5.groups.roles-ACT-before-reduction.target-short-side-native.median\endcsname{27}
\expandafter\def\csname pct@rev.desc.d5.groups.roles-ACT-before-reduction.target-short-side-native.median\endcsname{2720.0}
\expandafter\def\csname ppt@rev.desc.d5.groups.roles-ACT-before-reduction.target-short-side-native.median\endcsname{$+$2720.0}
\expandafter\def\csname val@rev.desc.d5.groups.roles-ACT-before-reduction.target-short-side-native.n\endcsname{943}
\expandafter\def\csname cnt@rev.desc.d5.groups.roles-ACT-before-reduction.target-short-side-native.n\endcsname{943}
\expandafter\def\csname pct@rev.desc.d5.groups.roles-ACT-before-reduction.target-short-side-native.n\endcsname{94300.0}
\expandafter\def\csname ppt@rev.desc.d5.groups.roles-ACT-before-reduction.target-short-side-native.n\endcsname{$+$94300.0}
\expandafter\def\csname val@rev.desc.d5.groups.roles-ACT-before-reduction.target-short-side-native.q25\endcsname{22.197}
\expandafter\def\csname cnt@rev.desc.d5.groups.roles-ACT-before-reduction.target-short-side-native.q25\endcsname{22}
\expandafter\def\csname pct@rev.desc.d5.groups.roles-ACT-before-reduction.target-short-side-native.q25\endcsname{2219.7}
\expandafter\def\csname ppt@rev.desc.d5.groups.roles-ACT-before-reduction.target-short-side-native.q25\endcsname{$+$2219.7}
\expandafter\def\csname val@rev.desc.d5.groups.roles-ACT-before-reduction.target-short-side-native.q75\endcsname{33.600}
\expandafter\def\csname cnt@rev.desc.d5.groups.roles-ACT-before-reduction.target-short-side-native.q75\endcsname{34}
\expandafter\def\csname pct@rev.desc.d5.groups.roles-ACT-before-reduction.target-short-side-native.q75\endcsname{3360.0}
\expandafter\def\csname ppt@rev.desc.d5.groups.roles-ACT-before-reduction.target-short-side-native.q75\endcsname{$+$3360.0}
\expandafter\def\csname val@rev.desc.d5.groups.roles-ACT-before-reduction.ui-category-share.Button\endcsname{0.531}
\expandafter\def\csname cnt@rev.desc.d5.groups.roles-ACT-before-reduction.ui-category-share.Button\endcsname{1}
\expandafter\def\csname pct@rev.desc.d5.groups.roles-ACT-before-reduction.ui-category-share.Button\endcsname{53.1}
\expandafter\def\csname ppt@rev.desc.d5.groups.roles-ACT-before-reduction.ui-category-share.Button\endcsname{$+$53.1}
\expandafter\def\csname val@rev.desc.d5.groups.roles-ACT-before-reduction.ui-category-share.Input Elements\endcsname{0.344}
\expandafter\def\csname cnt@rev.desc.d5.groups.roles-ACT-before-reduction.ui-category-share.Input Elements\endcsname{0}
\expandafter\def\csname pct@rev.desc.d5.groups.roles-ACT-before-reduction.ui-category-share.Input Elements\endcsname{34.4}
\expandafter\def\csname ppt@rev.desc.d5.groups.roles-ACT-before-reduction.ui-category-share.Input Elements\endcsname{$+$34.4}
\expandafter\def\csname val@rev.desc.d5.groups.roles-ACT-before-reduction.ui-category-share.Menu\endcsname{0.125}
\expandafter\def\csname cnt@rev.desc.d5.groups.roles-ACT-before-reduction.ui-category-share.Menu\endcsname{0}
\expandafter\def\csname pct@rev.desc.d5.groups.roles-ACT-before-reduction.ui-category-share.Menu\endcsname{12.5}
\expandafter\def\csname ppt@rev.desc.d5.groups.roles-ACT-before-reduction.ui-category-share.Menu\endcsname{$+$12.5}
\expandafter\def\csname val@rev.desc.d5.groups.used-A.element-type-n.icon\endcsname{41}
\expandafter\def\csname cnt@rev.desc.d5.groups.used-A.element-type-n.icon\endcsname{41}
\expandafter\def\csname pct@rev.desc.d5.groups.used-A.element-type-n.icon\endcsname{4100.0}
\expandafter\def\csname ppt@rev.desc.d5.groups.used-A.element-type-n.icon\endcsname{$+$4100.0}
\expandafter\def\csname val@rev.desc.d5.groups.used-A.element-type-n.text\endcsname{23}
\expandafter\def\csname cnt@rev.desc.d5.groups.used-A.element-type-n.text\endcsname{23}
\expandafter\def\csname pct@rev.desc.d5.groups.used-A.element-type-n.text\endcsname{2300.0}
\expandafter\def\csname ppt@rev.desc.d5.groups.used-A.element-type-n.text\endcsname{$+$2300.0}
\expandafter\def\csname val@rev.desc.d5.groups.used-A.element-type-share.icon\endcsname{0.641}
\expandafter\def\csname cnt@rev.desc.d5.groups.used-A.element-type-share.icon\endcsname{1}
\expandafter\def\csname pct@rev.desc.d5.groups.used-A.element-type-share.icon\endcsname{64.1}
\expandafter\def\csname ppt@rev.desc.d5.groups.used-A.element-type-share.icon\endcsname{$+$64.1}
\expandafter\def\csname val@rev.desc.d5.groups.used-A.element-type-share.text\endcsname{0.359}
\expandafter\def\csname cnt@rev.desc.d5.groups.used-A.element-type-share.text\endcsname{0}
\expandafter\def\csname pct@rev.desc.d5.groups.used-A.element-type-share.text\endcsname{35.9}
\expandafter\def\csname ppt@rev.desc.d5.groups.used-A.element-type-share.text\endcsname{$+$35.9}
\expandafter\def\csname val@rev.desc.d5.groups.used-A.image-long-edge.median\endcsname{1920.000}
\expandafter\def\csname cnt@rev.desc.d5.groups.used-A.image-long-edge.median\endcsname{1{,}920}
\expandafter\def\csname pct@rev.desc.d5.groups.used-A.image-long-edge.median\endcsname{192000.0}
\expandafter\def\csname ppt@rev.desc.d5.groups.used-A.image-long-edge.median\endcsname{$+$192000.0}
\expandafter\def\csname val@rev.desc.d5.groups.used-A.image-long-edge.n\endcsname{64}
\expandafter\def\csname cnt@rev.desc.d5.groups.used-A.image-long-edge.n\endcsname{64}
\expandafter\def\csname pct@rev.desc.d5.groups.used-A.image-long-edge.n\endcsname{6400.0}
\expandafter\def\csname ppt@rev.desc.d5.groups.used-A.image-long-edge.n\endcsname{$+$6400.0}
\expandafter\def\csname val@rev.desc.d5.groups.used-A.image-long-edge.q25\endcsname{1680.000}
\expandafter\def\csname cnt@rev.desc.d5.groups.used-A.image-long-edge.q25\endcsname{1{,}680}
\expandafter\def\csname pct@rev.desc.d5.groups.used-A.image-long-edge.q25\endcsname{168000.0}
\expandafter\def\csname ppt@rev.desc.d5.groups.used-A.image-long-edge.q25\endcsname{$+$168000.0}
\expandafter\def\csname val@rev.desc.d5.groups.used-A.image-long-edge.q75\endcsname{2310.000}
\expandafter\def\csname cnt@rev.desc.d5.groups.used-A.image-long-edge.q75\endcsname{2{,}310}
\expandafter\def\csname pct@rev.desc.d5.groups.used-A.image-long-edge.q75\endcsname{231000.0}
\expandafter\def\csname ppt@rev.desc.d5.groups.used-A.image-long-edge.q75\endcsname{$+$231000.0}
\expandafter\def\csname val@rev.desc.d5.groups.used-A.n\endcsname{64}
\expandafter\def\csname cnt@rev.desc.d5.groups.used-A.n\endcsname{64}
\expandafter\def\csname pct@rev.desc.d5.groups.used-A.n\endcsname{6400.0}
\expandafter\def\csname ppt@rev.desc.d5.groups.used-A.n\endcsname{$+$6400.0}
\expandafter\def\csname val@rev.desc.d5.groups.used-A.n-apps\endcsname{8}
\expandafter\def\csname cnt@rev.desc.d5.groups.used-A.n-apps\endcsname{8}
\expandafter\def\csname pct@rev.desc.d5.groups.used-A.n-apps\endcsname{800.0}
\expandafter\def\csname ppt@rev.desc.d5.groups.used-A.n-apps\endcsname{$+$800.0}
\expandafter\def\csname val@rev.desc.d5.groups.used-A.n-screenshots\endcsname{58}
\expandafter\def\csname cnt@rev.desc.d5.groups.used-A.n-screenshots\endcsname{58}
\expandafter\def\csname pct@rev.desc.d5.groups.used-A.n-screenshots\endcsname{5800.0}
\expandafter\def\csname ppt@rev.desc.d5.groups.used-A.n-screenshots\endcsname{$+$5800.0}
\expandafter\def\csname val@rev.desc.d5.groups.used-A.share-from-screenshots-with-gt2-roots\endcsname{0.031}
\expandafter\def\csname cnt@rev.desc.d5.groups.used-A.share-from-screenshots-with-gt2-roots\endcsname{0}
\expandafter\def\csname pct@rev.desc.d5.groups.used-A.share-from-screenshots-with-gt2-roots\endcsname{3.1}
\expandafter\def\csname ppt@rev.desc.d5.groups.used-A.share-from-screenshots-with-gt2-roots\endcsname{$+$3.1}
\expandafter\def\csname val@rev.desc.d5.groups.used-A.share-with-functional-prompt\endcsname{1.000}
\expandafter\def\csname cnt@rev.desc.d5.groups.used-A.share-with-functional-prompt\endcsname{1}
\expandafter\def\csname pct@rev.desc.d5.groups.used-A.share-with-functional-prompt\endcsname{100.0}
\expandafter\def\csname ppt@rev.desc.d5.groups.used-A.share-with-functional-prompt\endcsname{$+$100.0}
\expandafter\def\csname val@rev.desc.d5.groups.used-A.target-area-fraction.median\endcsname{0.001}
\expandafter\def\csname cnt@rev.desc.d5.groups.used-A.target-area-fraction.median\endcsname{0}
\expandafter\def\csname pct@rev.desc.d5.groups.used-A.target-area-fraction.median\endcsname{0.1}
\expandafter\def\csname ppt@rev.desc.d5.groups.used-A.target-area-fraction.median\endcsname{$+$0.1}
\expandafter\def\csname val@rev.desc.d5.groups.used-A.target-area-fraction.n\endcsname{64}
\expandafter\def\csname cnt@rev.desc.d5.groups.used-A.target-area-fraction.n\endcsname{64}
\expandafter\def\csname pct@rev.desc.d5.groups.used-A.target-area-fraction.n\endcsname{6400.0}
\expandafter\def\csname ppt@rev.desc.d5.groups.used-A.target-area-fraction.n\endcsname{$+$6400.0}
\expandafter\def\csname val@rev.desc.d5.groups.used-A.target-area-fraction.q25\endcsname{0.001}
\expandafter\def\csname cnt@rev.desc.d5.groups.used-A.target-area-fraction.q25\endcsname{0}
\expandafter\def\csname pct@rev.desc.d5.groups.used-A.target-area-fraction.q25\endcsname{0.1}
\expandafter\def\csname ppt@rev.desc.d5.groups.used-A.target-area-fraction.q25\endcsname{$+$0.1}
\expandafter\def\csname val@rev.desc.d5.groups.used-A.target-area-fraction.q75\endcsname{0.002}
\expandafter\def\csname cnt@rev.desc.d5.groups.used-A.target-area-fraction.q75\endcsname{0}
\expandafter\def\csname pct@rev.desc.d5.groups.used-A.target-area-fraction.q75\endcsname{0.2}
\expandafter\def\csname ppt@rev.desc.d5.groups.used-A.target-area-fraction.q75\endcsname{$+$0.2}
\expandafter\def\csname val@rev.desc.d5.groups.used-A.target-short-side-native.median\endcsname{30.849}
\expandafter\def\csname cnt@rev.desc.d5.groups.used-A.target-short-side-native.median\endcsname{31}
\expandafter\def\csname pct@rev.desc.d5.groups.used-A.target-short-side-native.median\endcsname{3084.9}
\expandafter\def\csname ppt@rev.desc.d5.groups.used-A.target-short-side-native.median\endcsname{$+$3084.9}
\expandafter\def\csname val@rev.desc.d5.groups.used-A.target-short-side-native.n\endcsname{64}
\expandafter\def\csname cnt@rev.desc.d5.groups.used-A.target-short-side-native.n\endcsname{64}
\expandafter\def\csname pct@rev.desc.d5.groups.used-A.target-short-side-native.n\endcsname{6400.0}
\expandafter\def\csname ppt@rev.desc.d5.groups.used-A.target-short-side-native.n\endcsname{$+$6400.0}
\expandafter\def\csname val@rev.desc.d5.groups.used-A.target-short-side-native.q25\endcsname{24.622}
\expandafter\def\csname cnt@rev.desc.d5.groups.used-A.target-short-side-native.q25\endcsname{25}
\expandafter\def\csname pct@rev.desc.d5.groups.used-A.target-short-side-native.q25\endcsname{2462.2}
\expandafter\def\csname ppt@rev.desc.d5.groups.used-A.target-short-side-native.q25\endcsname{$+$2462.2}
\expandafter\def\csname val@rev.desc.d5.groups.used-A.target-short-side-native.q75\endcsname{44.314}
\expandafter\def\csname cnt@rev.desc.d5.groups.used-A.target-short-side-native.q75\endcsname{44}
\expandafter\def\csname pct@rev.desc.d5.groups.used-A.target-short-side-native.q75\endcsname{4431.4}
\expandafter\def\csname ppt@rev.desc.d5.groups.used-A.target-short-side-native.q75\endcsname{$+$4431.4}
\expandafter\def\csname val@rev.desc.d5.groups.used-A.ui-category-share.Button\endcsname{0.375}
\expandafter\def\csname cnt@rev.desc.d5.groups.used-A.ui-category-share.Button\endcsname{0}
\expandafter\def\csname pct@rev.desc.d5.groups.used-A.ui-category-share.Button\endcsname{37.5}
\expandafter\def\csname ppt@rev.desc.d5.groups.used-A.ui-category-share.Button\endcsname{$+$37.5}
\expandafter\def\csname val@rev.desc.d5.groups.used-A.ui-category-share.Input Elements\endcsname{0.500}
\expandafter\def\csname cnt@rev.desc.d5.groups.used-A.ui-category-share.Input Elements\endcsname{0}
\expandafter\def\csname pct@rev.desc.d5.groups.used-A.ui-category-share.Input Elements\endcsname{50.0}
\expandafter\def\csname ppt@rev.desc.d5.groups.used-A.ui-category-share.Input Elements\endcsname{$+$50.0}
\expandafter\def\csname val@rev.desc.d5.groups.used-A.ui-category-share.Menu\endcsname{0.125}
\expandafter\def\csname cnt@rev.desc.d5.groups.used-A.ui-category-share.Menu\endcsname{0}
\expandafter\def\csname pct@rev.desc.d5.groups.used-A.ui-category-share.Menu\endcsname{12.5}
\expandafter\def\csname ppt@rev.desc.d5.groups.used-A.ui-category-share.Menu\endcsname{$+$12.5}
\expandafter\def\csname val@rev.desc.d5.groups.used-ACT.element-type-n.icon\endcsname{235}
\expandafter\def\csname cnt@rev.desc.d5.groups.used-ACT.element-type-n.icon\endcsname{235}
\expandafter\def\csname pct@rev.desc.d5.groups.used-ACT.element-type-n.icon\endcsname{23500.0}
\expandafter\def\csname ppt@rev.desc.d5.groups.used-ACT.element-type-n.icon\endcsname{$+$23500.0}
\expandafter\def\csname val@rev.desc.d5.groups.used-ACT.element-type-n.text\endcsname{69}
\expandafter\def\csname cnt@rev.desc.d5.groups.used-ACT.element-type-n.text\endcsname{69}
\expandafter\def\csname pct@rev.desc.d5.groups.used-ACT.element-type-n.text\endcsname{6900.0}
\expandafter\def\csname ppt@rev.desc.d5.groups.used-ACT.element-type-n.text\endcsname{$+$6900.0}
\expandafter\def\csname val@rev.desc.d5.groups.used-ACT.element-type-share.icon\endcsname{0.773}
\expandafter\def\csname cnt@rev.desc.d5.groups.used-ACT.element-type-share.icon\endcsname{1}
\expandafter\def\csname pct@rev.desc.d5.groups.used-ACT.element-type-share.icon\endcsname{77.3}
\expandafter\def\csname ppt@rev.desc.d5.groups.used-ACT.element-type-share.icon\endcsname{$+$77.3}
\expandafter\def\csname val@rev.desc.d5.groups.used-ACT.element-type-share.text\endcsname{0.227}
\expandafter\def\csname cnt@rev.desc.d5.groups.used-ACT.element-type-share.text\endcsname{0}
\expandafter\def\csname pct@rev.desc.d5.groups.used-ACT.element-type-share.text\endcsname{22.7}
\expandafter\def\csname ppt@rev.desc.d5.groups.used-ACT.element-type-share.text\endcsname{$+$22.7}
\expandafter\def\csname val@rev.desc.d5.groups.used-ACT.image-long-edge.median\endcsname{1920.000}
\expandafter\def\csname cnt@rev.desc.d5.groups.used-ACT.image-long-edge.median\endcsname{1{,}920}
\expandafter\def\csname pct@rev.desc.d5.groups.used-ACT.image-long-edge.median\endcsname{192000.0}
\expandafter\def\csname ppt@rev.desc.d5.groups.used-ACT.image-long-edge.median\endcsname{$+$192000.0}
\expandafter\def\csname val@rev.desc.d5.groups.used-ACT.image-long-edge.n\endcsname{304}
\expandafter\def\csname cnt@rev.desc.d5.groups.used-ACT.image-long-edge.n\endcsname{304}
\expandafter\def\csname pct@rev.desc.d5.groups.used-ACT.image-long-edge.n\endcsname{30400.0}
\expandafter\def\csname ppt@rev.desc.d5.groups.used-ACT.image-long-edge.n\endcsname{$+$30400.0}
\expandafter\def\csname val@rev.desc.d5.groups.used-ACT.image-long-edge.q25\endcsname{1680.000}
\expandafter\def\csname cnt@rev.desc.d5.groups.used-ACT.image-long-edge.q25\endcsname{1{,}680}
\expandafter\def\csname pct@rev.desc.d5.groups.used-ACT.image-long-edge.q25\endcsname{168000.0}
\expandafter\def\csname ppt@rev.desc.d5.groups.used-ACT.image-long-edge.q25\endcsname{$+$168000.0}
\expandafter\def\csname val@rev.desc.d5.groups.used-ACT.image-long-edge.q75\endcsname{1920.000}
\expandafter\def\csname cnt@rev.desc.d5.groups.used-ACT.image-long-edge.q75\endcsname{1{,}920}
\expandafter\def\csname pct@rev.desc.d5.groups.used-ACT.image-long-edge.q75\endcsname{192000.0}
\expandafter\def\csname ppt@rev.desc.d5.groups.used-ACT.image-long-edge.q75\endcsname{$+$192000.0}
\expandafter\def\csname val@rev.desc.d5.groups.used-ACT.n\endcsname{304}
\expandafter\def\csname cnt@rev.desc.d5.groups.used-ACT.n\endcsname{304}
\expandafter\def\csname pct@rev.desc.d5.groups.used-ACT.n\endcsname{30400.0}
\expandafter\def\csname ppt@rev.desc.d5.groups.used-ACT.n\endcsname{$+$30400.0}
\expandafter\def\csname val@rev.desc.d5.groups.used-ACT.n-apps\endcsname{44}
\expandafter\def\csname cnt@rev.desc.d5.groups.used-ACT.n-apps\endcsname{44}
\expandafter\def\csname pct@rev.desc.d5.groups.used-ACT.n-apps\endcsname{4400.0}
\expandafter\def\csname ppt@rev.desc.d5.groups.used-ACT.n-apps\endcsname{$+$4400.0}
\expandafter\def\csname val@rev.desc.d5.groups.used-ACT.n-screenshots\endcsname{281}
\expandafter\def\csname cnt@rev.desc.d5.groups.used-ACT.n-screenshots\endcsname{281}
\expandafter\def\csname pct@rev.desc.d5.groups.used-ACT.n-screenshots\endcsname{28100.0}
\expandafter\def\csname ppt@rev.desc.d5.groups.used-ACT.n-screenshots\endcsname{$+$28100.0}
\expandafter\def\csname val@rev.desc.d5.groups.used-ACT.share-from-screenshots-with-gt2-roots\endcsname{0.151}
\expandafter\def\csname cnt@rev.desc.d5.groups.used-ACT.share-from-screenshots-with-gt2-roots\endcsname{0}
\expandafter\def\csname pct@rev.desc.d5.groups.used-ACT.share-from-screenshots-with-gt2-roots\endcsname{15.1}
\expandafter\def\csname ppt@rev.desc.d5.groups.used-ACT.share-from-screenshots-with-gt2-roots\endcsname{$+$15.1}
\expandafter\def\csname val@rev.desc.d5.groups.used-ACT.share-with-functional-prompt\endcsname{1.000}
\expandafter\def\csname cnt@rev.desc.d5.groups.used-ACT.share-with-functional-prompt\endcsname{1}
\expandafter\def\csname pct@rev.desc.d5.groups.used-ACT.share-with-functional-prompt\endcsname{100.0}
\expandafter\def\csname ppt@rev.desc.d5.groups.used-ACT.share-with-functional-prompt\endcsname{$+$100.0}
\expandafter\def\csname val@rev.desc.d5.groups.used-ACT.target-area-fraction.median\endcsname{0.001}
\expandafter\def\csname cnt@rev.desc.d5.groups.used-ACT.target-area-fraction.median\endcsname{0}
\expandafter\def\csname pct@rev.desc.d5.groups.used-ACT.target-area-fraction.median\endcsname{0.1}
\expandafter\def\csname ppt@rev.desc.d5.groups.used-ACT.target-area-fraction.median\endcsname{$+$0.1}
\expandafter\def\csname val@rev.desc.d5.groups.used-ACT.target-area-fraction.n\endcsname{304}
\expandafter\def\csname cnt@rev.desc.d5.groups.used-ACT.target-area-fraction.n\endcsname{304}
\expandafter\def\csname pct@rev.desc.d5.groups.used-ACT.target-area-fraction.n\endcsname{30400.0}
\expandafter\def\csname ppt@rev.desc.d5.groups.used-ACT.target-area-fraction.n\endcsname{$+$30400.0}
\expandafter\def\csname val@rev.desc.d5.groups.used-ACT.target-area-fraction.q25\endcsname{0.000}
\expandafter\def\csname cnt@rev.desc.d5.groups.used-ACT.target-area-fraction.q25\endcsname{0}
\expandafter\def\csname pct@rev.desc.d5.groups.used-ACT.target-area-fraction.q25\endcsname{0.0}
\expandafter\def\csname ppt@rev.desc.d5.groups.used-ACT.target-area-fraction.q25\endcsname{0.0}
\expandafter\def\csname val@rev.desc.d5.groups.used-ACT.target-area-fraction.q75\endcsname{0.001}
\expandafter\def\csname cnt@rev.desc.d5.groups.used-ACT.target-area-fraction.q75\endcsname{0}
\expandafter\def\csname pct@rev.desc.d5.groups.used-ACT.target-area-fraction.q75\endcsname{0.1}
\expandafter\def\csname ppt@rev.desc.d5.groups.used-ACT.target-area-fraction.q75\endcsname{$+$0.1}
\expandafter\def\csname val@rev.desc.d5.groups.used-ACT.target-short-side-native.median\endcsname{27.771}
\expandafter\def\csname cnt@rev.desc.d5.groups.used-ACT.target-short-side-native.median\endcsname{28}
\expandafter\def\csname pct@rev.desc.d5.groups.used-ACT.target-short-side-native.median\endcsname{2777.1}
\expandafter\def\csname ppt@rev.desc.d5.groups.used-ACT.target-short-side-native.median\endcsname{$+$2777.1}
\expandafter\def\csname val@rev.desc.d5.groups.used-ACT.target-short-side-native.n\endcsname{304}
\expandafter\def\csname cnt@rev.desc.d5.groups.used-ACT.target-short-side-native.n\endcsname{304}
\expandafter\def\csname pct@rev.desc.d5.groups.used-ACT.target-short-side-native.n\endcsname{30400.0}
\expandafter\def\csname ppt@rev.desc.d5.groups.used-ACT.target-short-side-native.n\endcsname{$+$30400.0}
\expandafter\def\csname val@rev.desc.d5.groups.used-ACT.target-short-side-native.q25\endcsname{23.349}
\expandafter\def\csname cnt@rev.desc.d5.groups.used-ACT.target-short-side-native.q25\endcsname{23}
\expandafter\def\csname pct@rev.desc.d5.groups.used-ACT.target-short-side-native.q25\endcsname{2334.9}
\expandafter\def\csname ppt@rev.desc.d5.groups.used-ACT.target-short-side-native.q25\endcsname{$+$2334.9}
\expandafter\def\csname val@rev.desc.d5.groups.used-ACT.target-short-side-native.q75\endcsname{37.029}
\expandafter\def\csname cnt@rev.desc.d5.groups.used-ACT.target-short-side-native.q75\endcsname{37}
\expandafter\def\csname pct@rev.desc.d5.groups.used-ACT.target-short-side-native.q75\endcsname{3702.9}
\expandafter\def\csname ppt@rev.desc.d5.groups.used-ACT.target-short-side-native.q75\endcsname{$+$3702.9}
\expandafter\def\csname val@rev.desc.d5.groups.used-ACT.ui-category-share.Button\endcsname{0.500}
\expandafter\def\csname cnt@rev.desc.d5.groups.used-ACT.ui-category-share.Button\endcsname{0}
\expandafter\def\csname pct@rev.desc.d5.groups.used-ACT.ui-category-share.Button\endcsname{50.0}
\expandafter\def\csname ppt@rev.desc.d5.groups.used-ACT.ui-category-share.Button\endcsname{$+$50.0}
\expandafter\def\csname val@rev.desc.d5.groups.used-ACT.ui-category-share.Input Elements\endcsname{0.372}
\expandafter\def\csname cnt@rev.desc.d5.groups.used-ACT.ui-category-share.Input Elements\endcsname{0}
\expandafter\def\csname pct@rev.desc.d5.groups.used-ACT.ui-category-share.Input Elements\endcsname{37.2}
\expandafter\def\csname ppt@rev.desc.d5.groups.used-ACT.ui-category-share.Input Elements\endcsname{$+$37.2}
\expandafter\def\csname val@rev.desc.d5.groups.used-ACT.ui-category-share.Menu\endcsname{0.128}
\expandafter\def\csname cnt@rev.desc.d5.groups.used-ACT.ui-category-share.Menu\endcsname{0}
\expandafter\def\csname pct@rev.desc.d5.groups.used-ACT.ui-category-share.Menu\endcsname{12.8}
\expandafter\def\csname ppt@rev.desc.d5.groups.used-ACT.ui-category-share.Menu\endcsname{$+$12.8}
\expandafter\def\csname val@rev.desc.d5.groups.used-C.element-type-n.icon\endcsname{80}
\expandafter\def\csname cnt@rev.desc.d5.groups.used-C.element-type-n.icon\endcsname{80}
\expandafter\def\csname pct@rev.desc.d5.groups.used-C.element-type-n.icon\endcsname{8000.0}
\expandafter\def\csname ppt@rev.desc.d5.groups.used-C.element-type-n.icon\endcsname{$+$8000.0}
\expandafter\def\csname val@rev.desc.d5.groups.used-C.element-type-n.text\endcsname{16}
\expandafter\def\csname cnt@rev.desc.d5.groups.used-C.element-type-n.text\endcsname{16}
\expandafter\def\csname pct@rev.desc.d5.groups.used-C.element-type-n.text\endcsname{1600.0}
\expandafter\def\csname ppt@rev.desc.d5.groups.used-C.element-type-n.text\endcsname{$+$1600.0}
\expandafter\def\csname val@rev.desc.d5.groups.used-C.element-type-share.icon\endcsname{0.833}
\expandafter\def\csname cnt@rev.desc.d5.groups.used-C.element-type-share.icon\endcsname{1}
\expandafter\def\csname pct@rev.desc.d5.groups.used-C.element-type-share.icon\endcsname{83.3}
\expandafter\def\csname ppt@rev.desc.d5.groups.used-C.element-type-share.icon\endcsname{$+$83.3}
\expandafter\def\csname val@rev.desc.d5.groups.used-C.element-type-share.text\endcsname{0.167}
\expandafter\def\csname cnt@rev.desc.d5.groups.used-C.element-type-share.text\endcsname{0}
\expandafter\def\csname pct@rev.desc.d5.groups.used-C.element-type-share.text\endcsname{16.7}
\expandafter\def\csname ppt@rev.desc.d5.groups.used-C.element-type-share.text\endcsname{$+$16.7}
\expandafter\def\csname val@rev.desc.d5.groups.used-C.image-long-edge.median\endcsname{1920.000}
\expandafter\def\csname cnt@rev.desc.d5.groups.used-C.image-long-edge.median\endcsname{1{,}920}
\expandafter\def\csname pct@rev.desc.d5.groups.used-C.image-long-edge.median\endcsname{192000.0}
\expandafter\def\csname ppt@rev.desc.d5.groups.used-C.image-long-edge.median\endcsname{$+$192000.0}
\expandafter\def\csname val@rev.desc.d5.groups.used-C.image-long-edge.n\endcsname{96}
\expandafter\def\csname cnt@rev.desc.d5.groups.used-C.image-long-edge.n\endcsname{96}
\expandafter\def\csname pct@rev.desc.d5.groups.used-C.image-long-edge.n\endcsname{9600.0}
\expandafter\def\csname ppt@rev.desc.d5.groups.used-C.image-long-edge.n\endcsname{$+$9600.0}
\expandafter\def\csname val@rev.desc.d5.groups.used-C.image-long-edge.q25\endcsname{1773.000}
\expandafter\def\csname cnt@rev.desc.d5.groups.used-C.image-long-edge.q25\endcsname{1{,}773}
\expandafter\def\csname pct@rev.desc.d5.groups.used-C.image-long-edge.q25\endcsname{177300.0}
\expandafter\def\csname ppt@rev.desc.d5.groups.used-C.image-long-edge.q25\endcsname{$+$177300.0}
\expandafter\def\csname val@rev.desc.d5.groups.used-C.image-long-edge.q75\endcsname{1920.000}
\expandafter\def\csname cnt@rev.desc.d5.groups.used-C.image-long-edge.q75\endcsname{1{,}920}
\expandafter\def\csname pct@rev.desc.d5.groups.used-C.image-long-edge.q75\endcsname{192000.0}
\expandafter\def\csname ppt@rev.desc.d5.groups.used-C.image-long-edge.q75\endcsname{$+$192000.0}
\expandafter\def\csname val@rev.desc.d5.groups.used-C.n\endcsname{96}
\expandafter\def\csname cnt@rev.desc.d5.groups.used-C.n\endcsname{96}
\expandafter\def\csname pct@rev.desc.d5.groups.used-C.n\endcsname{9600.0}
\expandafter\def\csname ppt@rev.desc.d5.groups.used-C.n\endcsname{$+$9600.0}
\expandafter\def\csname val@rev.desc.d5.groups.used-C.n-apps\endcsname{12}
\expandafter\def\csname cnt@rev.desc.d5.groups.used-C.n-apps\endcsname{12}
\expandafter\def\csname pct@rev.desc.d5.groups.used-C.n-apps\endcsname{1200.0}
\expandafter\def\csname ppt@rev.desc.d5.groups.used-C.n-apps\endcsname{$+$1200.0}
\expandafter\def\csname val@rev.desc.d5.groups.used-C.n-screenshots\endcsname{87}
\expandafter\def\csname cnt@rev.desc.d5.groups.used-C.n-screenshots\endcsname{87}
\expandafter\def\csname pct@rev.desc.d5.groups.used-C.n-screenshots\endcsname{8700.0}
\expandafter\def\csname ppt@rev.desc.d5.groups.used-C.n-screenshots\endcsname{$+$8700.0}
\expandafter\def\csname val@rev.desc.d5.groups.used-C.share-from-screenshots-with-gt2-roots\endcsname{0.135}
\expandafter\def\csname cnt@rev.desc.d5.groups.used-C.share-from-screenshots-with-gt2-roots\endcsname{0}
\expandafter\def\csname pct@rev.desc.d5.groups.used-C.share-from-screenshots-with-gt2-roots\endcsname{13.5}
\expandafter\def\csname ppt@rev.desc.d5.groups.used-C.share-from-screenshots-with-gt2-roots\endcsname{$+$13.5}
\expandafter\def\csname val@rev.desc.d5.groups.used-C.share-with-functional-prompt\endcsname{1.000}
\expandafter\def\csname cnt@rev.desc.d5.groups.used-C.share-with-functional-prompt\endcsname{1}
\expandafter\def\csname pct@rev.desc.d5.groups.used-C.share-with-functional-prompt\endcsname{100.0}
\expandafter\def\csname ppt@rev.desc.d5.groups.used-C.share-with-functional-prompt\endcsname{$+$100.0}
\expandafter\def\csname val@rev.desc.d5.groups.used-C.target-area-fraction.median\endcsname{0.001}
\expandafter\def\csname cnt@rev.desc.d5.groups.used-C.target-area-fraction.median\endcsname{0}
\expandafter\def\csname pct@rev.desc.d5.groups.used-C.target-area-fraction.median\endcsname{0.1}
\expandafter\def\csname ppt@rev.desc.d5.groups.used-C.target-area-fraction.median\endcsname{$+$0.1}
\expandafter\def\csname val@rev.desc.d5.groups.used-C.target-area-fraction.n\endcsname{96}
\expandafter\def\csname cnt@rev.desc.d5.groups.used-C.target-area-fraction.n\endcsname{96}
\expandafter\def\csname pct@rev.desc.d5.groups.used-C.target-area-fraction.n\endcsname{9600.0}
\expandafter\def\csname ppt@rev.desc.d5.groups.used-C.target-area-fraction.n\endcsname{$+$9600.0}
\expandafter\def\csname val@rev.desc.d5.groups.used-C.target-area-fraction.q25\endcsname{0.000}
\expandafter\def\csname cnt@rev.desc.d5.groups.used-C.target-area-fraction.q25\endcsname{0}
\expandafter\def\csname pct@rev.desc.d5.groups.used-C.target-area-fraction.q25\endcsname{0.0}
\expandafter\def\csname ppt@rev.desc.d5.groups.used-C.target-area-fraction.q25\endcsname{0.0}
\expandafter\def\csname val@rev.desc.d5.groups.used-C.target-area-fraction.q75\endcsname{0.001}
\expandafter\def\csname cnt@rev.desc.d5.groups.used-C.target-area-fraction.q75\endcsname{0}
\expandafter\def\csname pct@rev.desc.d5.groups.used-C.target-area-fraction.q75\endcsname{0.1}
\expandafter\def\csname ppt@rev.desc.d5.groups.used-C.target-area-fraction.q75\endcsname{$+$0.1}
\expandafter\def\csname val@rev.desc.d5.groups.used-C.target-short-side-native.median\endcsname{26.596}
\expandafter\def\csname cnt@rev.desc.d5.groups.used-C.target-short-side-native.median\endcsname{27}
\expandafter\def\csname pct@rev.desc.d5.groups.used-C.target-short-side-native.median\endcsname{2659.6}
\expandafter\def\csname ppt@rev.desc.d5.groups.used-C.target-short-side-native.median\endcsname{$+$2659.6}
\expandafter\def\csname val@rev.desc.d5.groups.used-C.target-short-side-native.n\endcsname{96}
\expandafter\def\csname cnt@rev.desc.d5.groups.used-C.target-short-side-native.n\endcsname{96}
\expandafter\def\csname pct@rev.desc.d5.groups.used-C.target-short-side-native.n\endcsname{9600.0}
\expandafter\def\csname ppt@rev.desc.d5.groups.used-C.target-short-side-native.n\endcsname{$+$9600.0}
\expandafter\def\csname val@rev.desc.d5.groups.used-C.target-short-side-native.q25\endcsname{23.360}
\expandafter\def\csname cnt@rev.desc.d5.groups.used-C.target-short-side-native.q25\endcsname{23}
\expandafter\def\csname pct@rev.desc.d5.groups.used-C.target-short-side-native.q25\endcsname{2336.0}
\expandafter\def\csname ppt@rev.desc.d5.groups.used-C.target-short-side-native.q25\endcsname{$+$2336.0}
\expandafter\def\csname val@rev.desc.d5.groups.used-C.target-short-side-native.q75\endcsname{34.217}
\expandafter\def\csname cnt@rev.desc.d5.groups.used-C.target-short-side-native.q75\endcsname{34}
\expandafter\def\csname pct@rev.desc.d5.groups.used-C.target-short-side-native.q75\endcsname{3421.7}
\expandafter\def\csname ppt@rev.desc.d5.groups.used-C.target-short-side-native.q75\endcsname{$+$3421.7}
\expandafter\def\csname val@rev.desc.d5.groups.used-C.ui-category-share.Button\endcsname{0.490}
\expandafter\def\csname cnt@rev.desc.d5.groups.used-C.ui-category-share.Button\endcsname{0}
\expandafter\def\csname pct@rev.desc.d5.groups.used-C.ui-category-share.Button\endcsname{49.0}
\expandafter\def\csname ppt@rev.desc.d5.groups.used-C.ui-category-share.Button\endcsname{$+$49.0}
\expandafter\def\csname val@rev.desc.d5.groups.used-C.ui-category-share.Input Elements\endcsname{0.365}
\expandafter\def\csname cnt@rev.desc.d5.groups.used-C.ui-category-share.Input Elements\endcsname{0}
\expandafter\def\csname pct@rev.desc.d5.groups.used-C.ui-category-share.Input Elements\endcsname{36.5}
\expandafter\def\csname ppt@rev.desc.d5.groups.used-C.ui-category-share.Input Elements\endcsname{$+$36.5}
\expandafter\def\csname val@rev.desc.d5.groups.used-C.ui-category-share.Menu\endcsname{0.146}
\expandafter\def\csname cnt@rev.desc.d5.groups.used-C.ui-category-share.Menu\endcsname{0}
\expandafter\def\csname pct@rev.desc.d5.groups.used-C.ui-category-share.Menu\endcsname{14.6}
\expandafter\def\csname ppt@rev.desc.d5.groups.used-C.ui-category-share.Menu\endcsname{$+$14.6}
\expandafter\def\csname val@rev.desc.d5.groups.used-T.element-type-n.icon\endcsname{114}
\expandafter\def\csname cnt@rev.desc.d5.groups.used-T.element-type-n.icon\endcsname{114}
\expandafter\def\csname pct@rev.desc.d5.groups.used-T.element-type-n.icon\endcsname{11400.0}
\expandafter\def\csname ppt@rev.desc.d5.groups.used-T.element-type-n.icon\endcsname{$+$11400.0}
\expandafter\def\csname val@rev.desc.d5.groups.used-T.element-type-n.text\endcsname{30}
\expandafter\def\csname cnt@rev.desc.d5.groups.used-T.element-type-n.text\endcsname{30}
\expandafter\def\csname pct@rev.desc.d5.groups.used-T.element-type-n.text\endcsname{3000.0}
\expandafter\def\csname ppt@rev.desc.d5.groups.used-T.element-type-n.text\endcsname{$+$3000.0}
\expandafter\def\csname val@rev.desc.d5.groups.used-T.element-type-share.icon\endcsname{0.792}
\expandafter\def\csname cnt@rev.desc.d5.groups.used-T.element-type-share.icon\endcsname{1}
\expandafter\def\csname pct@rev.desc.d5.groups.used-T.element-type-share.icon\endcsname{79.2}
\expandafter\def\csname ppt@rev.desc.d5.groups.used-T.element-type-share.icon\endcsname{$+$79.2}
\expandafter\def\csname val@rev.desc.d5.groups.used-T.element-type-share.text\endcsname{0.208}
\expandafter\def\csname cnt@rev.desc.d5.groups.used-T.element-type-share.text\endcsname{0}
\expandafter\def\csname pct@rev.desc.d5.groups.used-T.element-type-share.text\endcsname{20.8}
\expandafter\def\csname ppt@rev.desc.d5.groups.used-T.element-type-share.text\endcsname{$+$20.8}
\expandafter\def\csname val@rev.desc.d5.groups.used-T.image-long-edge.median\endcsname{1920.000}
\expandafter\def\csname cnt@rev.desc.d5.groups.used-T.image-long-edge.median\endcsname{1{,}920}
\expandafter\def\csname pct@rev.desc.d5.groups.used-T.image-long-edge.median\endcsname{192000.0}
\expandafter\def\csname ppt@rev.desc.d5.groups.used-T.image-long-edge.median\endcsname{$+$192000.0}
\expandafter\def\csname val@rev.desc.d5.groups.used-T.image-long-edge.n\endcsname{144}
\expandafter\def\csname cnt@rev.desc.d5.groups.used-T.image-long-edge.n\endcsname{144}
\expandafter\def\csname pct@rev.desc.d5.groups.used-T.image-long-edge.n\endcsname{14400.0}
\expandafter\def\csname ppt@rev.desc.d5.groups.used-T.image-long-edge.n\endcsname{$+$14400.0}
\expandafter\def\csname val@rev.desc.d5.groups.used-T.image-long-edge.q25\endcsname{1680.000}
\expandafter\def\csname cnt@rev.desc.d5.groups.used-T.image-long-edge.q25\endcsname{1{,}680}
\expandafter\def\csname pct@rev.desc.d5.groups.used-T.image-long-edge.q25\endcsname{168000.0}
\expandafter\def\csname ppt@rev.desc.d5.groups.used-T.image-long-edge.q25\endcsname{$+$168000.0}
\expandafter\def\csname val@rev.desc.d5.groups.used-T.image-long-edge.q75\endcsname{1920.000}
\expandafter\def\csname cnt@rev.desc.d5.groups.used-T.image-long-edge.q75\endcsname{1{,}920}
\expandafter\def\csname pct@rev.desc.d5.groups.used-T.image-long-edge.q75\endcsname{192000.0}
\expandafter\def\csname ppt@rev.desc.d5.groups.used-T.image-long-edge.q75\endcsname{$+$192000.0}
\expandafter\def\csname val@rev.desc.d5.groups.used-T.n\endcsname{144}
\expandafter\def\csname cnt@rev.desc.d5.groups.used-T.n\endcsname{144}
\expandafter\def\csname pct@rev.desc.d5.groups.used-T.n\endcsname{14400.0}
\expandafter\def\csname ppt@rev.desc.d5.groups.used-T.n\endcsname{$+$14400.0}
\expandafter\def\csname val@rev.desc.d5.groups.used-T.n-apps\endcsname{24}
\expandafter\def\csname cnt@rev.desc.d5.groups.used-T.n-apps\endcsname{24}
\expandafter\def\csname pct@rev.desc.d5.groups.used-T.n-apps\endcsname{2400.0}
\expandafter\def\csname ppt@rev.desc.d5.groups.used-T.n-apps\endcsname{$+$2400.0}
\expandafter\def\csname val@rev.desc.d5.groups.used-T.n-screenshots\endcsname{136}
\expandafter\def\csname cnt@rev.desc.d5.groups.used-T.n-screenshots\endcsname{136}
\expandafter\def\csname pct@rev.desc.d5.groups.used-T.n-screenshots\endcsname{13600.0}
\expandafter\def\csname ppt@rev.desc.d5.groups.used-T.n-screenshots\endcsname{$+$13600.0}
\expandafter\def\csname val@rev.desc.d5.groups.used-T.share-from-screenshots-with-gt2-roots\endcsname{0.215}
\expandafter\def\csname cnt@rev.desc.d5.groups.used-T.share-from-screenshots-with-gt2-roots\endcsname{0}
\expandafter\def\csname pct@rev.desc.d5.groups.used-T.share-from-screenshots-with-gt2-roots\endcsname{21.5}
\expandafter\def\csname ppt@rev.desc.d5.groups.used-T.share-from-screenshots-with-gt2-roots\endcsname{$+$21.5}
\expandafter\def\csname val@rev.desc.d5.groups.used-T.share-with-functional-prompt\endcsname{1.000}
\expandafter\def\csname cnt@rev.desc.d5.groups.used-T.share-with-functional-prompt\endcsname{1}
\expandafter\def\csname pct@rev.desc.d5.groups.used-T.share-with-functional-prompt\endcsname{100.0}
\expandafter\def\csname ppt@rev.desc.d5.groups.used-T.share-with-functional-prompt\endcsname{$+$100.0}
\expandafter\def\csname val@rev.desc.d5.groups.used-T.target-area-fraction.median\endcsname{0.001}
\expandafter\def\csname cnt@rev.desc.d5.groups.used-T.target-area-fraction.median\endcsname{0}
\expandafter\def\csname pct@rev.desc.d5.groups.used-T.target-area-fraction.median\endcsname{0.1}
\expandafter\def\csname ppt@rev.desc.d5.groups.used-T.target-area-fraction.median\endcsname{$+$0.1}
\expandafter\def\csname val@rev.desc.d5.groups.used-T.target-area-fraction.n\endcsname{144}
\expandafter\def\csname cnt@rev.desc.d5.groups.used-T.target-area-fraction.n\endcsname{144}
\expandafter\def\csname pct@rev.desc.d5.groups.used-T.target-area-fraction.n\endcsname{14400.0}
\expandafter\def\csname ppt@rev.desc.d5.groups.used-T.target-area-fraction.n\endcsname{$+$14400.0}
\expandafter\def\csname val@rev.desc.d5.groups.used-T.target-area-fraction.q25\endcsname{0.000}
\expandafter\def\csname cnt@rev.desc.d5.groups.used-T.target-area-fraction.q25\endcsname{0}
\expandafter\def\csname pct@rev.desc.d5.groups.used-T.target-area-fraction.q25\endcsname{0.0}
\expandafter\def\csname ppt@rev.desc.d5.groups.used-T.target-area-fraction.q25\endcsname{0.0}
\expandafter\def\csname val@rev.desc.d5.groups.used-T.target-area-fraction.q75\endcsname{0.001}
\expandafter\def\csname cnt@rev.desc.d5.groups.used-T.target-area-fraction.q75\endcsname{0}
\expandafter\def\csname pct@rev.desc.d5.groups.used-T.target-area-fraction.q75\endcsname{0.1}
\expandafter\def\csname ppt@rev.desc.d5.groups.used-T.target-area-fraction.q75\endcsname{$+$0.1}
\expandafter\def\csname val@rev.desc.d5.groups.used-T.target-short-side-native.median\endcsname{27.593}
\expandafter\def\csname cnt@rev.desc.d5.groups.used-T.target-short-side-native.median\endcsname{28}
\expandafter\def\csname pct@rev.desc.d5.groups.used-T.target-short-side-native.median\endcsname{2759.3}
\expandafter\def\csname ppt@rev.desc.d5.groups.used-T.target-short-side-native.median\endcsname{$+$2759.3}
\expandafter\def\csname val@rev.desc.d5.groups.used-T.target-short-side-native.n\endcsname{144}
\expandafter\def\csname cnt@rev.desc.d5.groups.used-T.target-short-side-native.n\endcsname{144}
\expandafter\def\csname pct@rev.desc.d5.groups.used-T.target-short-side-native.n\endcsname{14400.0}
\expandafter\def\csname ppt@rev.desc.d5.groups.used-T.target-short-side-native.n\endcsname{$+$14400.0}
\expandafter\def\csname val@rev.desc.d5.groups.used-T.target-short-side-native.q25\endcsname{22.919}
\expandafter\def\csname cnt@rev.desc.d5.groups.used-T.target-short-side-native.q25\endcsname{23}
\expandafter\def\csname pct@rev.desc.d5.groups.used-T.target-short-side-native.q25\endcsname{2291.9}
\expandafter\def\csname ppt@rev.desc.d5.groups.used-T.target-short-side-native.q25\endcsname{$+$2291.9}
\expandafter\def\csname val@rev.desc.d5.groups.used-T.target-short-side-native.q75\endcsname{36.000}
\expandafter\def\csname cnt@rev.desc.d5.groups.used-T.target-short-side-native.q75\endcsname{36}
\expandafter\def\csname pct@rev.desc.d5.groups.used-T.target-short-side-native.q75\endcsname{3600.0}
\expandafter\def\csname ppt@rev.desc.d5.groups.used-T.target-short-side-native.q75\endcsname{$+$3600.0}
\expandafter\def\csname val@rev.desc.d5.groups.used-T.ui-category-share.Button\endcsname{0.562}
\expandafter\def\csname cnt@rev.desc.d5.groups.used-T.ui-category-share.Button\endcsname{1}
\expandafter\def\csname pct@rev.desc.d5.groups.used-T.ui-category-share.Button\endcsname{56.2}
\expandafter\def\csname ppt@rev.desc.d5.groups.used-T.ui-category-share.Button\endcsname{$+$56.2}
\expandafter\def\csname val@rev.desc.d5.groups.used-T.ui-category-share.Input Elements\endcsname{0.319}
\expandafter\def\csname cnt@rev.desc.d5.groups.used-T.ui-category-share.Input Elements\endcsname{0}
\expandafter\def\csname pct@rev.desc.d5.groups.used-T.ui-category-share.Input Elements\endcsname{31.9}
\expandafter\def\csname ppt@rev.desc.d5.groups.used-T.ui-category-share.Input Elements\endcsname{$+$31.9}
\expandafter\def\csname val@rev.desc.d5.groups.used-T.ui-category-share.Menu\endcsname{0.118}
\expandafter\def\csname cnt@rev.desc.d5.groups.used-T.ui-category-share.Menu\endcsname{0}
\expandafter\def\csname pct@rev.desc.d5.groups.used-T.ui-category-share.Menu\endcsname{11.8}
\expandafter\def\csname ppt@rev.desc.d5.groups.used-T.ui-category-share.Menu\endcsname{$+$11.8}
\expandafter\def\csname val@rev.desc.d5.kept-in-pool\endcsname{1561}
\expandafter\def\csname cnt@rev.desc.d5.kept-in-pool\endcsname{1{,}561}
\expandafter\def\csname pct@rev.desc.d5.kept-in-pool\endcsname{156100.0}
\expandafter\def\csname ppt@rev.desc.d5.kept-in-pool\endcsname{$+$156100.0}
\expandafter\def\csname val@rev.desc.d5.kept-total-manifest\endcsname{1561}
\expandafter\def\csname cnt@rev.desc.d5.kept-total-manifest\endcsname{1{,}561}
\expandafter\def\csname pct@rev.desc.d5.kept-total-manifest\endcsname{156100.0}
\expandafter\def\csname ppt@rev.desc.d5.kept-total-manifest\endcsname{$+$156100.0}
\expandafter\def\csname val@rev.desc.d5.n-removed-with-functional-prompt\endcsname{183}
\expandafter\def\csname cnt@rev.desc.d5.n-removed-with-functional-prompt\endcsname{183}
\expandafter\def\csname pct@rev.desc.d5.n-removed-with-functional-prompt\endcsname{18300.0}
\expandafter\def\csname ppt@rev.desc.d5.n-removed-with-functional-prompt\endcsname{$+$18300.0}
\expandafter\def\csname val@rev.desc.d5.screenshots-in-pool\endcsname{1164}
\expandafter\def\csname cnt@rev.desc.d5.screenshots-in-pool\endcsname{1{,}164}
\expandafter\def\csname pct@rev.desc.d5.screenshots-in-pool\endcsname{116400.0}
\expandafter\def\csname ppt@rev.desc.d5.screenshots-in-pool\endcsname{$+$116400.0}
\expandafter\def\csname val@rev.desc.d5.screenshots-with-gt2-roots\endcsname{129}
\expandafter\def\csname cnt@rev.desc.d5.screenshots-with-gt2-roots\endcsname{129}
\expandafter\def\csname pct@rev.desc.d5.screenshots-with-gt2-roots\endcsname{12900.0}
\expandafter\def\csname ppt@rev.desc.d5.screenshots-with-gt2-roots\endcsname{$+$12900.0}
\expandafter\def\csname val@rev.desc.d5.sspro.full-benchmark.category-share.CAD\endcsname{0.165}
\expandafter\def\csname cnt@rev.desc.d5.sspro.full-benchmark.category-share.CAD\endcsname{0}
\expandafter\def\csname pct@rev.desc.d5.sspro.full-benchmark.category-share.CAD\endcsname{16.5}
\expandafter\def\csname ppt@rev.desc.d5.sspro.full-benchmark.category-share.CAD\endcsname{$+$16.5}
\expandafter\def\csname val@rev.desc.d5.sspro.full-benchmark.category-share.Creative\endcsname{0.216}
\expandafter\def\csname cnt@rev.desc.d5.sspro.full-benchmark.category-share.Creative\endcsname{0}
\expandafter\def\csname pct@rev.desc.d5.sspro.full-benchmark.category-share.Creative\endcsname{21.6}
\expandafter\def\csname ppt@rev.desc.d5.sspro.full-benchmark.category-share.Creative\endcsname{$+$21.6}
\expandafter\def\csname val@rev.desc.d5.sspro.full-benchmark.category-share.Dev\endcsname{0.189}
\expandafter\def\csname cnt@rev.desc.d5.sspro.full-benchmark.category-share.Dev\endcsname{0}
\expandafter\def\csname pct@rev.desc.d5.sspro.full-benchmark.category-share.Dev\endcsname{18.9}
\expandafter\def\csname ppt@rev.desc.d5.sspro.full-benchmark.category-share.Dev\endcsname{$+$18.9}
\expandafter\def\csname val@rev.desc.d5.sspro.full-benchmark.category-share.OS\endcsname{0.124}
\expandafter\def\csname cnt@rev.desc.d5.sspro.full-benchmark.category-share.OS\endcsname{0}
\expandafter\def\csname pct@rev.desc.d5.sspro.full-benchmark.category-share.OS\endcsname{12.4}
\expandafter\def\csname ppt@rev.desc.d5.sspro.full-benchmark.category-share.OS\endcsname{$+$12.4}
\expandafter\def\csname val@rev.desc.d5.sspro.full-benchmark.category-share.Office\endcsname{0.145}
\expandafter\def\csname cnt@rev.desc.d5.sspro.full-benchmark.category-share.Office\endcsname{0}
\expandafter\def\csname pct@rev.desc.d5.sspro.full-benchmark.category-share.Office\endcsname{14.5}
\expandafter\def\csname ppt@rev.desc.d5.sspro.full-benchmark.category-share.Office\endcsname{$+$14.5}
\expandafter\def\csname val@rev.desc.d5.sspro.full-benchmark.category-share.Scientific\endcsname{0.161}
\expandafter\def\csname cnt@rev.desc.d5.sspro.full-benchmark.category-share.Scientific\endcsname{0}
\expandafter\def\csname pct@rev.desc.d5.sspro.full-benchmark.category-share.Scientific\endcsname{16.1}
\expandafter\def\csname ppt@rev.desc.d5.sspro.full-benchmark.category-share.Scientific\endcsname{$+$16.1}
\expandafter\def\csname val@rev.desc.d5.sspro.full-benchmark.image-long-edge.median\endcsname{3456.000}
\expandafter\def\csname cnt@rev.desc.d5.sspro.full-benchmark.image-long-edge.median\endcsname{3{,}456}
\expandafter\def\csname pct@rev.desc.d5.sspro.full-benchmark.image-long-edge.median\endcsname{345600.0}
\expandafter\def\csname ppt@rev.desc.d5.sspro.full-benchmark.image-long-edge.median\endcsname{$+$345600.0}
\expandafter\def\csname val@rev.desc.d5.sspro.full-benchmark.image-long-edge.n\endcsname{1581}
\expandafter\def\csname cnt@rev.desc.d5.sspro.full-benchmark.image-long-edge.n\endcsname{1{,}581}
\expandafter\def\csname pct@rev.desc.d5.sspro.full-benchmark.image-long-edge.n\endcsname{158100.0}
\expandafter\def\csname ppt@rev.desc.d5.sspro.full-benchmark.image-long-edge.n\endcsname{$+$158100.0}
\expandafter\def\csname val@rev.desc.d5.sspro.full-benchmark.image-long-edge.q25\endcsname{2560.000}
\expandafter\def\csname cnt@rev.desc.d5.sspro.full-benchmark.image-long-edge.q25\endcsname{2{,}560}
\expandafter\def\csname pct@rev.desc.d5.sspro.full-benchmark.image-long-edge.q25\endcsname{256000.0}
\expandafter\def\csname ppt@rev.desc.d5.sspro.full-benchmark.image-long-edge.q25\endcsname{$+$256000.0}
\expandafter\def\csname val@rev.desc.d5.sspro.full-benchmark.image-long-edge.q75\endcsname{3840.000}
\expandafter\def\csname cnt@rev.desc.d5.sspro.full-benchmark.image-long-edge.q75\endcsname{3{,}840}
\expandafter\def\csname pct@rev.desc.d5.sspro.full-benchmark.image-long-edge.q75\endcsname{384000.0}
\expandafter\def\csname ppt@rev.desc.d5.sspro.full-benchmark.image-long-edge.q75\endcsname{$+$384000.0}
\expandafter\def\csname val@rev.desc.d5.sspro.full-benchmark.n\endcsname{1581}
\expandafter\def\csname cnt@rev.desc.d5.sspro.full-benchmark.n\endcsname{1{,}581}
\expandafter\def\csname pct@rev.desc.d5.sspro.full-benchmark.n\endcsname{158100.0}
\expandafter\def\csname ppt@rev.desc.d5.sspro.full-benchmark.n\endcsname{$+$158100.0}
\expandafter\def\csname val@rev.desc.d5.sspro.full-benchmark.n-apps\endcsname{26}
\expandafter\def\csname cnt@rev.desc.d5.sspro.full-benchmark.n-apps\endcsname{26}
\expandafter\def\csname pct@rev.desc.d5.sspro.full-benchmark.n-apps\endcsname{2600.0}
\expandafter\def\csname ppt@rev.desc.d5.sspro.full-benchmark.n-apps\endcsname{$+$2600.0}
\expandafter\def\csname val@rev.desc.d5.sspro.full-benchmark.target-short-side-native.median\endcsname{24.000}
\expandafter\def\csname cnt@rev.desc.d5.sspro.full-benchmark.target-short-side-native.median\endcsname{24}
\expandafter\def\csname pct@rev.desc.d5.sspro.full-benchmark.target-short-side-native.median\endcsname{2400.0}
\expandafter\def\csname ppt@rev.desc.d5.sspro.full-benchmark.target-short-side-native.median\endcsname{$+$2400.0}
\expandafter\def\csname val@rev.desc.d5.sspro.full-benchmark.target-short-side-native.n\endcsname{1581}
\expandafter\def\csname cnt@rev.desc.d5.sspro.full-benchmark.target-short-side-native.n\endcsname{1{,}581}
\expandafter\def\csname pct@rev.desc.d5.sspro.full-benchmark.target-short-side-native.n\endcsname{158100.0}
\expandafter\def\csname ppt@rev.desc.d5.sspro.full-benchmark.target-short-side-native.n\endcsname{$+$158100.0}
\expandafter\def\csname val@rev.desc.d5.sspro.full-benchmark.target-short-side-native.q25\endcsname{20.000}
\expandafter\def\csname cnt@rev.desc.d5.sspro.full-benchmark.target-short-side-native.q25\endcsname{20}
\expandafter\def\csname pct@rev.desc.d5.sspro.full-benchmark.target-short-side-native.q25\endcsname{2000.0}
\expandafter\def\csname ppt@rev.desc.d5.sspro.full-benchmark.target-short-side-native.q25\endcsname{$+$2000.0}
\expandafter\def\csname val@rev.desc.d5.sspro.full-benchmark.target-short-side-native.q75\endcsname{32.000}
\expandafter\def\csname cnt@rev.desc.d5.sspro.full-benchmark.target-short-side-native.q75\endcsname{32}
\expandafter\def\csname pct@rev.desc.d5.sspro.full-benchmark.target-short-side-native.q75\endcsname{3200.0}
\expandafter\def\csname ppt@rev.desc.d5.sspro.full-benchmark.target-short-side-native.q75\endcsname{$+$3200.0}
\expandafter\def\csname val@rev.desc.d5.sspro.full-benchmark.ui-type-share.icon\endcsname{0.382}
\expandafter\def\csname cnt@rev.desc.d5.sspro.full-benchmark.ui-type-share.icon\endcsname{0}
\expandafter\def\csname pct@rev.desc.d5.sspro.full-benchmark.ui-type-share.icon\endcsname{38.2}
\expandafter\def\csname ppt@rev.desc.d5.sspro.full-benchmark.ui-type-share.icon\endcsname{$+$38.2}
\expandafter\def\csname val@rev.desc.d5.sspro.full-benchmark.ui-type-share.text\endcsname{0.618}
\expandafter\def\csname cnt@rev.desc.d5.sspro.full-benchmark.ui-type-share.text\endcsname{1}
\expandafter\def\csname pct@rev.desc.d5.sspro.full-benchmark.ui-type-share.text\endcsname{61.8}
\expandafter\def\csname ppt@rev.desc.d5.sspro.full-benchmark.ui-type-share.text\endcsname{$+$61.8}
\expandafter\def\csname val@rev.desc.d5.sspro.used-ACT.category-share.CAD\endcsname{0.136}
\expandafter\def\csname cnt@rev.desc.d5.sspro.used-ACT.category-share.CAD\endcsname{0}
\expandafter\def\csname pct@rev.desc.d5.sspro.used-ACT.category-share.CAD\endcsname{13.6}
\expandafter\def\csname ppt@rev.desc.d5.sspro.used-ACT.category-share.CAD\endcsname{$+$13.6}
\expandafter\def\csname val@rev.desc.d5.sspro.used-ACT.category-share.Creative\endcsname{0.227}
\expandafter\def\csname cnt@rev.desc.d5.sspro.used-ACT.category-share.Creative\endcsname{0}
\expandafter\def\csname pct@rev.desc.d5.sspro.used-ACT.category-share.Creative\endcsname{22.7}
\expandafter\def\csname ppt@rev.desc.d5.sspro.used-ACT.category-share.Creative\endcsname{$+$22.7}
\expandafter\def\csname val@rev.desc.d5.sspro.used-ACT.category-share.Dev\endcsname{0.227}
\expandafter\def\csname cnt@rev.desc.d5.sspro.used-ACT.category-share.Dev\endcsname{0}
\expandafter\def\csname pct@rev.desc.d5.sspro.used-ACT.category-share.Dev\endcsname{22.7}
\expandafter\def\csname ppt@rev.desc.d5.sspro.used-ACT.category-share.Dev\endcsname{$+$22.7}
\expandafter\def\csname val@rev.desc.d5.sspro.used-ACT.category-share.OS\endcsname{0.136}
\expandafter\def\csname cnt@rev.desc.d5.sspro.used-ACT.category-share.OS\endcsname{0}
\expandafter\def\csname pct@rev.desc.d5.sspro.used-ACT.category-share.OS\endcsname{13.6}
\expandafter\def\csname ppt@rev.desc.d5.sspro.used-ACT.category-share.OS\endcsname{$+$13.6}
\expandafter\def\csname val@rev.desc.d5.sspro.used-ACT.category-share.Office\endcsname{0.136}
\expandafter\def\csname cnt@rev.desc.d5.sspro.used-ACT.category-share.Office\endcsname{0}
\expandafter\def\csname pct@rev.desc.d5.sspro.used-ACT.category-share.Office\endcsname{13.6}
\expandafter\def\csname ppt@rev.desc.d5.sspro.used-ACT.category-share.Office\endcsname{$+$13.6}
\expandafter\def\csname val@rev.desc.d5.sspro.used-ACT.category-share.Scientific\endcsname{0.136}
\expandafter\def\csname cnt@rev.desc.d5.sspro.used-ACT.category-share.Scientific\endcsname{0}
\expandafter\def\csname pct@rev.desc.d5.sspro.used-ACT.category-share.Scientific\endcsname{13.6}
\expandafter\def\csname ppt@rev.desc.d5.sspro.used-ACT.category-share.Scientific\endcsname{$+$13.6}
\expandafter\def\csname val@rev.desc.d5.sspro.used-ACT.image-long-edge.median\endcsname{3456.000}
\expandafter\def\csname cnt@rev.desc.d5.sspro.used-ACT.image-long-edge.median\endcsname{3{,}456}
\expandafter\def\csname pct@rev.desc.d5.sspro.used-ACT.image-long-edge.median\endcsname{345600.0}
\expandafter\def\csname ppt@rev.desc.d5.sspro.used-ACT.image-long-edge.median\endcsname{$+$345600.0}
\expandafter\def\csname val@rev.desc.d5.sspro.used-ACT.image-long-edge.n\endcsname{176}
\expandafter\def\csname cnt@rev.desc.d5.sspro.used-ACT.image-long-edge.n\endcsname{176}
\expandafter\def\csname pct@rev.desc.d5.sspro.used-ACT.image-long-edge.n\endcsname{17600.0}
\expandafter\def\csname ppt@rev.desc.d5.sspro.used-ACT.image-long-edge.n\endcsname{$+$17600.0}
\expandafter\def\csname val@rev.desc.d5.sspro.used-ACT.image-long-edge.q25\endcsname{2560.000}
\expandafter\def\csname cnt@rev.desc.d5.sspro.used-ACT.image-long-edge.q25\endcsname{2{,}560}
\expandafter\def\csname pct@rev.desc.d5.sspro.used-ACT.image-long-edge.q25\endcsname{256000.0}
\expandafter\def\csname ppt@rev.desc.d5.sspro.used-ACT.image-long-edge.q25\endcsname{$+$256000.0}
\expandafter\def\csname val@rev.desc.d5.sspro.used-ACT.image-long-edge.q75\endcsname{3840.000}
\expandafter\def\csname cnt@rev.desc.d5.sspro.used-ACT.image-long-edge.q75\endcsname{3{,}840}
\expandafter\def\csname pct@rev.desc.d5.sspro.used-ACT.image-long-edge.q75\endcsname{384000.0}
\expandafter\def\csname ppt@rev.desc.d5.sspro.used-ACT.image-long-edge.q75\endcsname{$+$384000.0}
\expandafter\def\csname val@rev.desc.d5.sspro.used-ACT.n\endcsname{176}
\expandafter\def\csname cnt@rev.desc.d5.sspro.used-ACT.n\endcsname{176}
\expandafter\def\csname pct@rev.desc.d5.sspro.used-ACT.n\endcsname{17600.0}
\expandafter\def\csname ppt@rev.desc.d5.sspro.used-ACT.n\endcsname{$+$17600.0}
\expandafter\def\csname val@rev.desc.d5.sspro.used-ACT.n-apps\endcsname{22}
\expandafter\def\csname cnt@rev.desc.d5.sspro.used-ACT.n-apps\endcsname{22}
\expandafter\def\csname pct@rev.desc.d5.sspro.used-ACT.n-apps\endcsname{2200.0}
\expandafter\def\csname ppt@rev.desc.d5.sspro.used-ACT.n-apps\endcsname{$+$2200.0}
\expandafter\def\csname val@rev.desc.d5.sspro.used-ACT.target-short-side-native.median\endcsname{23.000}
\expandafter\def\csname cnt@rev.desc.d5.sspro.used-ACT.target-short-side-native.median\endcsname{23}
\expandafter\def\csname pct@rev.desc.d5.sspro.used-ACT.target-short-side-native.median\endcsname{2300.0}
\expandafter\def\csname ppt@rev.desc.d5.sspro.used-ACT.target-short-side-native.median\endcsname{$+$2300.0}
\expandafter\def\csname val@rev.desc.d5.sspro.used-ACT.target-short-side-native.n\endcsname{176}
\expandafter\def\csname cnt@rev.desc.d5.sspro.used-ACT.target-short-side-native.n\endcsname{176}
\expandafter\def\csname pct@rev.desc.d5.sspro.used-ACT.target-short-side-native.n\endcsname{17600.0}
\expandafter\def\csname ppt@rev.desc.d5.sspro.used-ACT.target-short-side-native.n\endcsname{$+$17600.0}
\expandafter\def\csname val@rev.desc.d5.sspro.used-ACT.target-short-side-native.q25\endcsname{19.000}
\expandafter\def\csname cnt@rev.desc.d5.sspro.used-ACT.target-short-side-native.q25\endcsname{19}
\expandafter\def\csname pct@rev.desc.d5.sspro.used-ACT.target-short-side-native.q25\endcsname{1900.0}
\expandafter\def\csname ppt@rev.desc.d5.sspro.used-ACT.target-short-side-native.q25\endcsname{$+$1900.0}
\expandafter\def\csname val@rev.desc.d5.sspro.used-ACT.target-short-side-native.q75\endcsname{30.000}
\expandafter\def\csname cnt@rev.desc.d5.sspro.used-ACT.target-short-side-native.q75\endcsname{30}
\expandafter\def\csname pct@rev.desc.d5.sspro.used-ACT.target-short-side-native.q75\endcsname{3000.0}
\expandafter\def\csname ppt@rev.desc.d5.sspro.used-ACT.target-short-side-native.q75\endcsname{$+$3000.0}
\expandafter\def\csname val@rev.desc.d5.sspro.used-ACT.ui-type-share.icon\endcsname{0.398}
\expandafter\def\csname cnt@rev.desc.d5.sspro.used-ACT.ui-type-share.icon\endcsname{0}
\expandafter\def\csname pct@rev.desc.d5.sspro.used-ACT.ui-type-share.icon\endcsname{39.8}
\expandafter\def\csname ppt@rev.desc.d5.sspro.used-ACT.ui-type-share.icon\endcsname{$+$39.8}
\expandafter\def\csname val@rev.desc.d5.sspro.used-ACT.ui-type-share.text\endcsname{0.602}
\expandafter\def\csname cnt@rev.desc.d5.sspro.used-ACT.ui-type-share.text\endcsname{1}
\expandafter\def\csname pct@rev.desc.d5.sspro.used-ACT.ui-type-share.text\endcsname{60.2}
\expandafter\def\csname ppt@rev.desc.d5.sspro.used-ACT.ui-type-share.text\endcsname{$+$60.2}
\expandafter\def\csname val@rev.desc.d5.unique-valid-roots\endcsname{1744}
\expandafter\def\csname cnt@rev.desc.d5.unique-valid-roots\endcsname{1{,}744}
\expandafter\def\csname pct@rev.desc.d5.unique-valid-roots\endcsname{174400.0}
\expandafter\def\csname ppt@rev.desc.d5.unique-valid-roots\endcsname{$+$174400.0}
\expandafter\def\csname val@rev.desc.d6.sspro.p1.C.box.catalog-pooled.N-agree-stable\endcsname{109}
\expandafter\def\csname cnt@rev.desc.d6.sspro.p1.C.box.catalog-pooled.N-agree-stable\endcsname{109}
\expandafter\def\csname pct@rev.desc.d6.sspro.p1.C.box.catalog-pooled.N-agree-stable\endcsname{10900.0}
\expandafter\def\csname ppt@rev.desc.d6.sspro.p1.C.box.catalog-pooled.N-agree-stable\endcsname{$+$10900.0}
\expandafter\def\csname val@rev.desc.d6.sspro.p1.C.box.catalog-pooled.N-disagree\endcsname{74}
\expandafter\def\csname cnt@rev.desc.d6.sspro.p1.C.box.catalog-pooled.N-disagree\endcsname{74}
\expandafter\def\csname pct@rev.desc.d6.sspro.p1.C.box.catalog-pooled.N-disagree\endcsname{7400.0}
\expandafter\def\csname ppt@rev.desc.d6.sspro.p1.C.box.catalog-pooled.N-disagree\endcsname{$+$7400.0}
\expandafter\def\csname val@rev.desc.d6.sspro.p1.C.box.catalog-pooled.boot-resid.ci95.0\endcsname{-0.033}
\expandafter\def\csname cnt@rev.desc.d6.sspro.p1.C.box.catalog-pooled.boot-resid.ci95.0\endcsname{0}
\expandafter\def\csname pct@rev.desc.d6.sspro.p1.C.box.catalog-pooled.boot-resid.ci95.0\endcsname{$-$3.3}
\expandafter\def\csname ppt@rev.desc.d6.sspro.p1.C.box.catalog-pooled.boot-resid.ci95.0\endcsname{$-$3.3}
\expandafter\def\csname val@rev.desc.d6.sspro.p1.C.box.catalog-pooled.boot-resid.ci95.1\endcsname{0.000}
\expandafter\def\csname cnt@rev.desc.d6.sspro.p1.C.box.catalog-pooled.boot-resid.ci95.1\endcsname{0}
\expandafter\def\csname pct@rev.desc.d6.sspro.p1.C.box.catalog-pooled.boot-resid.ci95.1\endcsname{0.0}
\expandafter\def\csname ppt@rev.desc.d6.sspro.p1.C.box.catalog-pooled.boot-resid.ci95.1\endcsname{0.0}
\expandafter\def\csname val@rev.desc.d6.sspro.p1.C.box.catalog-pooled.boot-resid.n-apps\endcsname{6}
\expandafter\def\csname cnt@rev.desc.d6.sspro.p1.C.box.catalog-pooled.boot-resid.n-apps\endcsname{6}
\expandafter\def\csname pct@rev.desc.d6.sspro.p1.C.box.catalog-pooled.boot-resid.n-apps\endcsname{600.0}
\expandafter\def\csname ppt@rev.desc.d6.sspro.p1.C.box.catalog-pooled.boot-resid.n-apps\endcsname{$+$600.0}
\expandafter\def\csname val@rev.desc.d6.sspro.p1.C.box.catalog-pooled.boot-resid.n-boot-valid\endcsname{4000}
\expandafter\def\csname cnt@rev.desc.d6.sspro.p1.C.box.catalog-pooled.boot-resid.n-boot-valid\endcsname{4{,}000}
\expandafter\def\csname pct@rev.desc.d6.sspro.p1.C.box.catalog-pooled.boot-resid.n-boot-valid\endcsname{400000.0}
\expandafter\def\csname ppt@rev.desc.d6.sspro.p1.C.box.catalog-pooled.boot-resid.n-boot-valid\endcsname{$+$400000.0}
\expandafter\def\csname val@rev.desc.d6.sspro.p1.C.box.catalog-pooled.boot-resid.value\endcsname{-0.009}
\expandafter\def\csname cnt@rev.desc.d6.sspro.p1.C.box.catalog-pooled.boot-resid.value\endcsname{0}
\expandafter\def\csname pct@rev.desc.d6.sspro.p1.C.box.catalog-pooled.boot-resid.value\endcsname{$-$0.9}
\expandafter\def\csname ppt@rev.desc.d6.sspro.p1.C.box.catalog-pooled.boot-resid.value\endcsname{$-$0.9}
\expandafter\def\csname val@rev.desc.d6.sspro.p1.C.box.catalog-pooled.boot-share.ci95.0\endcsname{0.359}
\expandafter\def\csname cnt@rev.desc.d6.sspro.p1.C.box.catalog-pooled.boot-share.ci95.0\endcsname{0}
\expandafter\def\csname pct@rev.desc.d6.sspro.p1.C.box.catalog-pooled.boot-share.ci95.0\endcsname{35.9}
\expandafter\def\csname ppt@rev.desc.d6.sspro.p1.C.box.catalog-pooled.boot-share.ci95.0\endcsname{$+$35.9}
\expandafter\def\csname val@rev.desc.d6.sspro.p1.C.box.catalog-pooled.boot-share.ci95.1\endcsname{0.750}
\expandafter\def\csname cnt@rev.desc.d6.sspro.p1.C.box.catalog-pooled.boot-share.ci95.1\endcsname{1}
\expandafter\def\csname pct@rev.desc.d6.sspro.p1.C.box.catalog-pooled.boot-share.ci95.1\endcsname{75.0}
\expandafter\def\csname ppt@rev.desc.d6.sspro.p1.C.box.catalog-pooled.boot-share.ci95.1\endcsname{$+$75.0}
\expandafter\def\csname val@rev.desc.d6.sspro.p1.C.box.catalog-pooled.boot-share.n-apps\endcsname{6}
\expandafter\def\csname cnt@rev.desc.d6.sspro.p1.C.box.catalog-pooled.boot-share.n-apps\endcsname{6}
\expandafter\def\csname pct@rev.desc.d6.sspro.p1.C.box.catalog-pooled.boot-share.n-apps\endcsname{600.0}
\expandafter\def\csname ppt@rev.desc.d6.sspro.p1.C.box.catalog-pooled.boot-share.n-apps\endcsname{$+$600.0}
\expandafter\def\csname val@rev.desc.d6.sspro.p1.C.box.catalog-pooled.boot-share.n-boot-valid\endcsname{4000}
\expandafter\def\csname cnt@rev.desc.d6.sspro.p1.C.box.catalog-pooled.boot-share.n-boot-valid\endcsname{4{,}000}
\expandafter\def\csname pct@rev.desc.d6.sspro.p1.C.box.catalog-pooled.boot-share.n-boot-valid\endcsname{400000.0}
\expandafter\def\csname ppt@rev.desc.d6.sspro.p1.C.box.catalog-pooled.boot-share.n-boot-valid\endcsname{$+$400000.0}
\expandafter\def\csname val@rev.desc.d6.sspro.p1.C.box.catalog-pooled.boot-share.value\endcsname{0.568}
\expandafter\def\csname cnt@rev.desc.d6.sspro.p1.C.box.catalog-pooled.boot-share.value\endcsname{1}
\expandafter\def\csname pct@rev.desc.d6.sspro.p1.C.box.catalog-pooled.boot-share.value\endcsname{56.8}
\expandafter\def\csname ppt@rev.desc.d6.sspro.p1.C.box.catalog-pooled.boot-share.value\endcsname{$+$56.8}
\expandafter\def\csname val@rev.desc.d6.sspro.p1.C.box.catalog-pooled.gap\endcsname{0.077}
\expandafter\def\csname cnt@rev.desc.d6.sspro.p1.C.box.catalog-pooled.gap\endcsname{0}
\expandafter\def\csname pct@rev.desc.d6.sspro.p1.C.box.catalog-pooled.gap\endcsname{7.7}
\expandafter\def\csname ppt@rev.desc.d6.sspro.p1.C.box.catalog-pooled.gap\endcsname{$+$7.7}
\expandafter\def\csname val@rev.desc.d6.sspro.p1.C.box.catalog-pooled.n\endcsname{192}
\expandafter\def\csname cnt@rev.desc.d6.sspro.p1.C.box.catalog-pooled.n\endcsname{192}
\expandafter\def\csname pct@rev.desc.d6.sspro.p1.C.box.catalog-pooled.n\endcsname{19200.0}
\expandafter\def\csname ppt@rev.desc.d6.sspro.p1.C.box.catalog-pooled.n\endcsname{$+$19200.0}
\expandafter\def\csname val@rev.desc.d6.sspro.p1.C.box.catalog-pooled.resid-agree-stable\endcsname{-0.009}
\expandafter\def\csname cnt@rev.desc.d6.sspro.p1.C.box.catalog-pooled.resid-agree-stable\endcsname{0}
\expandafter\def\csname pct@rev.desc.d6.sspro.p1.C.box.catalog-pooled.resid-agree-stable\endcsname{$-$0.9}
\expandafter\def\csname ppt@rev.desc.d6.sspro.p1.C.box.catalog-pooled.resid-agree-stable\endcsname{$-$0.9}
\expandafter\def\csname val@rev.desc.d6.sspro.p1.C.box.catalog-pooled.resid-disagree\endcsname{0.068}
\expandafter\def\csname cnt@rev.desc.d6.sspro.p1.C.box.catalog-pooled.resid-disagree\endcsname{0}
\expandafter\def\csname pct@rev.desc.d6.sspro.p1.C.box.catalog-pooled.resid-disagree\endcsname{6.8}
\expandafter\def\csname ppt@rev.desc.d6.sspro.p1.C.box.catalog-pooled.resid-disagree\endcsname{$+$6.8}
\expandafter\def\csname val@rev.desc.d6.sspro.p1.C.box.catalog-pooled.share-agree-stable\endcsname{0.568}
\expandafter\def\csname cnt@rev.desc.d6.sspro.p1.C.box.catalog-pooled.share-agree-stable\endcsname{1}
\expandafter\def\csname pct@rev.desc.d6.sspro.p1.C.box.catalog-pooled.share-agree-stable\endcsname{56.8}
\expandafter\def\csname ppt@rev.desc.d6.sspro.p1.C.box.catalog-pooled.share-agree-stable\endcsname{$+$56.8}
\expandafter\def\csname val@rev.desc.d6.sspro.p1.C.box.center-d1.N-agree-stable\endcsname{29}
\expandafter\def\csname cnt@rev.desc.d6.sspro.p1.C.box.center-d1.N-agree-stable\endcsname{29}
\expandafter\def\csname pct@rev.desc.d6.sspro.p1.C.box.center-d1.N-agree-stable\endcsname{2900.0}
\expandafter\def\csname ppt@rev.desc.d6.sspro.p1.C.box.center-d1.N-agree-stable\endcsname{$+$2900.0}
\expandafter\def\csname val@rev.desc.d6.sspro.p1.C.box.center-d1.N-disagree\endcsname{18}
\expandafter\def\csname cnt@rev.desc.d6.sspro.p1.C.box.center-d1.N-disagree\endcsname{18}
\expandafter\def\csname pct@rev.desc.d6.sspro.p1.C.box.center-d1.N-disagree\endcsname{1800.0}
\expandafter\def\csname ppt@rev.desc.d6.sspro.p1.C.box.center-d1.N-disagree\endcsname{$+$1800.0}
\expandafter\def\csname val@rev.desc.d6.sspro.p1.C.box.center-d1.boot-resid.ci95.0\endcsname{0.000}
\expandafter\def\csname cnt@rev.desc.d6.sspro.p1.C.box.center-d1.boot-resid.ci95.0\endcsname{0}
\expandafter\def\csname pct@rev.desc.d6.sspro.p1.C.box.center-d1.boot-resid.ci95.0\endcsname{0.0}
\expandafter\def\csname ppt@rev.desc.d6.sspro.p1.C.box.center-d1.boot-resid.ci95.0\endcsname{0.0}
\expandafter\def\csname val@rev.desc.d6.sspro.p1.C.box.center-d1.boot-resid.ci95.1\endcsname{0.000}
\expandafter\def\csname cnt@rev.desc.d6.sspro.p1.C.box.center-d1.boot-resid.ci95.1\endcsname{0}
\expandafter\def\csname pct@rev.desc.d6.sspro.p1.C.box.center-d1.boot-resid.ci95.1\endcsname{0.0}
\expandafter\def\csname ppt@rev.desc.d6.sspro.p1.C.box.center-d1.boot-resid.ci95.1\endcsname{0.0}
\expandafter\def\csname val@rev.desc.d6.sspro.p1.C.box.center-d1.boot-resid.n-apps\endcsname{6}
\expandafter\def\csname cnt@rev.desc.d6.sspro.p1.C.box.center-d1.boot-resid.n-apps\endcsname{6}
\expandafter\def\csname pct@rev.desc.d6.sspro.p1.C.box.center-d1.boot-resid.n-apps\endcsname{600.0}
\expandafter\def\csname ppt@rev.desc.d6.sspro.p1.C.box.center-d1.boot-resid.n-apps\endcsname{$+$600.0}
\expandafter\def\csname val@rev.desc.d6.sspro.p1.C.box.center-d1.boot-resid.n-boot-valid\endcsname{4000}
\expandafter\def\csname cnt@rev.desc.d6.sspro.p1.C.box.center-d1.boot-resid.n-boot-valid\endcsname{4{,}000}
\expandafter\def\csname pct@rev.desc.d6.sspro.p1.C.box.center-d1.boot-resid.n-boot-valid\endcsname{400000.0}
\expandafter\def\csname ppt@rev.desc.d6.sspro.p1.C.box.center-d1.boot-resid.n-boot-valid\endcsname{$+$400000.0}
\expandafter\def\csname val@rev.desc.d6.sspro.p1.C.box.center-d1.boot-resid.value\endcsname{0.000}
\expandafter\def\csname cnt@rev.desc.d6.sspro.p1.C.box.center-d1.boot-resid.value\endcsname{0}
\expandafter\def\csname pct@rev.desc.d6.sspro.p1.C.box.center-d1.boot-resid.value\endcsname{0.0}
\expandafter\def\csname ppt@rev.desc.d6.sspro.p1.C.box.center-d1.boot-resid.value\endcsname{0.0}
\expandafter\def\csname val@rev.desc.d6.sspro.p1.C.box.center-d1.boot-share.ci95.0\endcsname{0.375}
\expandafter\def\csname cnt@rev.desc.d6.sspro.p1.C.box.center-d1.boot-share.ci95.0\endcsname{0}
\expandafter\def\csname pct@rev.desc.d6.sspro.p1.C.box.center-d1.boot-share.ci95.0\endcsname{37.5}
\expandafter\def\csname ppt@rev.desc.d6.sspro.p1.C.box.center-d1.boot-share.ci95.0\endcsname{$+$37.5}
\expandafter\def\csname val@rev.desc.d6.sspro.p1.C.box.center-d1.boot-share.ci95.1\endcsname{0.792}
\expandafter\def\csname cnt@rev.desc.d6.sspro.p1.C.box.center-d1.boot-share.ci95.1\endcsname{1}
\expandafter\def\csname pct@rev.desc.d6.sspro.p1.C.box.center-d1.boot-share.ci95.1\endcsname{79.2}
\expandafter\def\csname ppt@rev.desc.d6.sspro.p1.C.box.center-d1.boot-share.ci95.1\endcsname{$+$79.2}
\expandafter\def\csname val@rev.desc.d6.sspro.p1.C.box.center-d1.boot-share.n-apps\endcsname{6}
\expandafter\def\csname cnt@rev.desc.d6.sspro.p1.C.box.center-d1.boot-share.n-apps\endcsname{6}
\expandafter\def\csname pct@rev.desc.d6.sspro.p1.C.box.center-d1.boot-share.n-apps\endcsname{600.0}
\expandafter\def\csname ppt@rev.desc.d6.sspro.p1.C.box.center-d1.boot-share.n-apps\endcsname{$+$600.0}
\expandafter\def\csname val@rev.desc.d6.sspro.p1.C.box.center-d1.boot-share.n-boot-valid\endcsname{4000}
\expandafter\def\csname cnt@rev.desc.d6.sspro.p1.C.box.center-d1.boot-share.n-boot-valid\endcsname{4{,}000}
\expandafter\def\csname pct@rev.desc.d6.sspro.p1.C.box.center-d1.boot-share.n-boot-valid\endcsname{400000.0}
\expandafter\def\csname ppt@rev.desc.d6.sspro.p1.C.box.center-d1.boot-share.n-boot-valid\endcsname{$+$400000.0}
\expandafter\def\csname val@rev.desc.d6.sspro.p1.C.box.center-d1.boot-share.value\endcsname{0.604}
\expandafter\def\csname cnt@rev.desc.d6.sspro.p1.C.box.center-d1.boot-share.value\endcsname{1}
\expandafter\def\csname pct@rev.desc.d6.sspro.p1.C.box.center-d1.boot-share.value\endcsname{60.4}
\expandafter\def\csname ppt@rev.desc.d6.sspro.p1.C.box.center-d1.boot-share.value\endcsname{$+$60.4}
\expandafter\def\csname val@rev.desc.d6.sspro.p1.C.box.center-d1.n\endcsname{48}
\expandafter\def\csname cnt@rev.desc.d6.sspro.p1.C.box.center-d1.n\endcsname{48}
\expandafter\def\csname pct@rev.desc.d6.sspro.p1.C.box.center-d1.n\endcsname{4800.0}
\expandafter\def\csname ppt@rev.desc.d6.sspro.p1.C.box.center-d1.n\endcsname{$+$4800.0}
\expandafter\def\csname val@rev.desc.d6.sspro.p1.C.box.center-d1.resid-agree-stable\endcsname{0.000}
\expandafter\def\csname cnt@rev.desc.d6.sspro.p1.C.box.center-d1.resid-agree-stable\endcsname{0}
\expandafter\def\csname pct@rev.desc.d6.sspro.p1.C.box.center-d1.resid-agree-stable\endcsname{0.0}
\expandafter\def\csname ppt@rev.desc.d6.sspro.p1.C.box.center-d1.resid-agree-stable\endcsname{0.0}
\expandafter\def\csname val@rev.desc.d6.sspro.p1.C.box.center-d1.resid-disagree\endcsname{0.056}
\expandafter\def\csname cnt@rev.desc.d6.sspro.p1.C.box.center-d1.resid-disagree\endcsname{0}
\expandafter\def\csname pct@rev.desc.d6.sspro.p1.C.box.center-d1.resid-disagree\endcsname{5.6}
\expandafter\def\csname ppt@rev.desc.d6.sspro.p1.C.box.center-d1.resid-disagree\endcsname{$+$5.6}
\expandafter\def\csname val@rev.desc.d6.sspro.p1.C.box.center-d1.share-agree-stable\endcsname{0.604}
\expandafter\def\csname cnt@rev.desc.d6.sspro.p1.C.box.center-d1.share-agree-stable\endcsname{1}
\expandafter\def\csname pct@rev.desc.d6.sspro.p1.C.box.center-d1.share-agree-stable\endcsname{60.4}
\expandafter\def\csname ppt@rev.desc.d6.sspro.p1.C.box.center-d1.share-agree-stable\endcsname{$+$60.4}
\expandafter\def\csname val@rev.desc.d6.sspro.p1.C.box.context-d1.N-agree-stable\endcsname{27}
\expandafter\def\csname cnt@rev.desc.d6.sspro.p1.C.box.context-d1.N-agree-stable\endcsname{27}
\expandafter\def\csname pct@rev.desc.d6.sspro.p1.C.box.context-d1.N-agree-stable\endcsname{2700.0}
\expandafter\def\csname ppt@rev.desc.d6.sspro.p1.C.box.context-d1.N-agree-stable\endcsname{$+$2700.0}
\expandafter\def\csname val@rev.desc.d6.sspro.p1.C.box.context-d1.N-disagree\endcsname{18}
\expandafter\def\csname cnt@rev.desc.d6.sspro.p1.C.box.context-d1.N-disagree\endcsname{18}
\expandafter\def\csname pct@rev.desc.d6.sspro.p1.C.box.context-d1.N-disagree\endcsname{1800.0}
\expandafter\def\csname ppt@rev.desc.d6.sspro.p1.C.box.context-d1.N-disagree\endcsname{$+$1800.0}
\expandafter\def\csname val@rev.desc.d6.sspro.p1.C.box.context-d1.boot-resid.ci95.0\endcsname{0.000}
\expandafter\def\csname cnt@rev.desc.d6.sspro.p1.C.box.context-d1.boot-resid.ci95.0\endcsname{0}
\expandafter\def\csname pct@rev.desc.d6.sspro.p1.C.box.context-d1.boot-resid.ci95.0\endcsname{0.0}
\expandafter\def\csname ppt@rev.desc.d6.sspro.p1.C.box.context-d1.boot-resid.ci95.0\endcsname{0.0}
\expandafter\def\csname val@rev.desc.d6.sspro.p1.C.box.context-d1.boot-resid.ci95.1\endcsname{0.000}
\expandafter\def\csname cnt@rev.desc.d6.sspro.p1.C.box.context-d1.boot-resid.ci95.1\endcsname{0}
\expandafter\def\csname pct@rev.desc.d6.sspro.p1.C.box.context-d1.boot-resid.ci95.1\endcsname{0.0}
\expandafter\def\csname ppt@rev.desc.d6.sspro.p1.C.box.context-d1.boot-resid.ci95.1\endcsname{0.0}
\expandafter\def\csname val@rev.desc.d6.sspro.p1.C.box.context-d1.boot-resid.n-apps\endcsname{6}
\expandafter\def\csname cnt@rev.desc.d6.sspro.p1.C.box.context-d1.boot-resid.n-apps\endcsname{6}
\expandafter\def\csname pct@rev.desc.d6.sspro.p1.C.box.context-d1.boot-resid.n-apps\endcsname{600.0}
\expandafter\def\csname ppt@rev.desc.d6.sspro.p1.C.box.context-d1.boot-resid.n-apps\endcsname{$+$600.0}
\expandafter\def\csname val@rev.desc.d6.sspro.p1.C.box.context-d1.boot-resid.n-boot-valid\endcsname{4000}
\expandafter\def\csname cnt@rev.desc.d6.sspro.p1.C.box.context-d1.boot-resid.n-boot-valid\endcsname{4{,}000}
\expandafter\def\csname pct@rev.desc.d6.sspro.p1.C.box.context-d1.boot-resid.n-boot-valid\endcsname{400000.0}
\expandafter\def\csname ppt@rev.desc.d6.sspro.p1.C.box.context-d1.boot-resid.n-boot-valid\endcsname{$+$400000.0}
\expandafter\def\csname val@rev.desc.d6.sspro.p1.C.box.context-d1.boot-resid.value\endcsname{0.000}
\expandafter\def\csname cnt@rev.desc.d6.sspro.p1.C.box.context-d1.boot-resid.value\endcsname{0}
\expandafter\def\csname pct@rev.desc.d6.sspro.p1.C.box.context-d1.boot-resid.value\endcsname{0.0}
\expandafter\def\csname ppt@rev.desc.d6.sspro.p1.C.box.context-d1.boot-resid.value\endcsname{0.0}
\expandafter\def\csname val@rev.desc.d6.sspro.p1.C.box.context-d1.boot-share.ci95.0\endcsname{0.354}
\expandafter\def\csname cnt@rev.desc.d6.sspro.p1.C.box.context-d1.boot-share.ci95.0\endcsname{0}
\expandafter\def\csname pct@rev.desc.d6.sspro.p1.C.box.context-d1.boot-share.ci95.0\endcsname{35.4}
\expandafter\def\csname ppt@rev.desc.d6.sspro.p1.C.box.context-d1.boot-share.ci95.0\endcsname{$+$35.4}
\expandafter\def\csname val@rev.desc.d6.sspro.p1.C.box.context-d1.boot-share.ci95.1\endcsname{0.771}
\expandafter\def\csname cnt@rev.desc.d6.sspro.p1.C.box.context-d1.boot-share.ci95.1\endcsname{1}
\expandafter\def\csname pct@rev.desc.d6.sspro.p1.C.box.context-d1.boot-share.ci95.1\endcsname{77.1}
\expandafter\def\csname ppt@rev.desc.d6.sspro.p1.C.box.context-d1.boot-share.ci95.1\endcsname{$+$77.1}
\expandafter\def\csname val@rev.desc.d6.sspro.p1.C.box.context-d1.boot-share.n-apps\endcsname{6}
\expandafter\def\csname cnt@rev.desc.d6.sspro.p1.C.box.context-d1.boot-share.n-apps\endcsname{6}
\expandafter\def\csname pct@rev.desc.d6.sspro.p1.C.box.context-d1.boot-share.n-apps\endcsname{600.0}
\expandafter\def\csname ppt@rev.desc.d6.sspro.p1.C.box.context-d1.boot-share.n-apps\endcsname{$+$600.0}
\expandafter\def\csname val@rev.desc.d6.sspro.p1.C.box.context-d1.boot-share.n-boot-valid\endcsname{4000}
\expandafter\def\csname cnt@rev.desc.d6.sspro.p1.C.box.context-d1.boot-share.n-boot-valid\endcsname{4{,}000}
\expandafter\def\csname pct@rev.desc.d6.sspro.p1.C.box.context-d1.boot-share.n-boot-valid\endcsname{400000.0}
\expandafter\def\csname ppt@rev.desc.d6.sspro.p1.C.box.context-d1.boot-share.n-boot-valid\endcsname{$+$400000.0}
\expandafter\def\csname val@rev.desc.d6.sspro.p1.C.box.context-d1.boot-share.value\endcsname{0.562}
\expandafter\def\csname cnt@rev.desc.d6.sspro.p1.C.box.context-d1.boot-share.value\endcsname{1}
\expandafter\def\csname pct@rev.desc.d6.sspro.p1.C.box.context-d1.boot-share.value\endcsname{56.2}
\expandafter\def\csname ppt@rev.desc.d6.sspro.p1.C.box.context-d1.boot-share.value\endcsname{$+$56.2}
\expandafter\def\csname val@rev.desc.d6.sspro.p1.C.box.context-d1.n\endcsname{48}
\expandafter\def\csname cnt@rev.desc.d6.sspro.p1.C.box.context-d1.n\endcsname{48}
\expandafter\def\csname pct@rev.desc.d6.sspro.p1.C.box.context-d1.n\endcsname{4800.0}
\expandafter\def\csname ppt@rev.desc.d6.sspro.p1.C.box.context-d1.n\endcsname{$+$4800.0}
\expandafter\def\csname val@rev.desc.d6.sspro.p1.C.box.context-d1.resid-agree-stable\endcsname{0.000}
\expandafter\def\csname cnt@rev.desc.d6.sspro.p1.C.box.context-d1.resid-agree-stable\endcsname{0}
\expandafter\def\csname pct@rev.desc.d6.sspro.p1.C.box.context-d1.resid-agree-stable\endcsname{0.0}
\expandafter\def\csname ppt@rev.desc.d6.sspro.p1.C.box.context-d1.resid-agree-stable\endcsname{0.0}
\expandafter\def\csname val@rev.desc.d6.sspro.p1.C.box.context-d1.resid-disagree\endcsname{0.000}
\expandafter\def\csname cnt@rev.desc.d6.sspro.p1.C.box.context-d1.resid-disagree\endcsname{0}
\expandafter\def\csname pct@rev.desc.d6.sspro.p1.C.box.context-d1.resid-disagree\endcsname{0.0}
\expandafter\def\csname ppt@rev.desc.d6.sspro.p1.C.box.context-d1.resid-disagree\endcsname{0.0}
\expandafter\def\csname val@rev.desc.d6.sspro.p1.C.box.context-d1.share-agree-stable\endcsname{0.562}
\expandafter\def\csname cnt@rev.desc.d6.sspro.p1.C.box.context-d1.share-agree-stable\endcsname{1}
\expandafter\def\csname pct@rev.desc.d6.sspro.p1.C.box.context-d1.share-agree-stable\endcsname{56.2}
\expandafter\def\csname ppt@rev.desc.d6.sspro.p1.C.box.context-d1.share-agree-stable\endcsname{$+$56.2}
\expandafter\def\csname val@rev.desc.d6.sspro.p1.C.box.dual-d2.N-agree-stable\endcsname{25}
\expandafter\def\csname cnt@rev.desc.d6.sspro.p1.C.box.dual-d2.N-agree-stable\endcsname{25}
\expandafter\def\csname pct@rev.desc.d6.sspro.p1.C.box.dual-d2.N-agree-stable\endcsname{2500.0}
\expandafter\def\csname ppt@rev.desc.d6.sspro.p1.C.box.dual-d2.N-agree-stable\endcsname{$+$2500.0}
\expandafter\def\csname val@rev.desc.d6.sspro.p1.C.box.dual-d2.N-disagree\endcsname{20}
\expandafter\def\csname cnt@rev.desc.d6.sspro.p1.C.box.dual-d2.N-disagree\endcsname{20}
\expandafter\def\csname pct@rev.desc.d6.sspro.p1.C.box.dual-d2.N-disagree\endcsname{2000.0}
\expandafter\def\csname ppt@rev.desc.d6.sspro.p1.C.box.dual-d2.N-disagree\endcsname{$+$2000.0}
\expandafter\def\csname val@rev.desc.d6.sspro.p1.C.box.dual-d2.boot-resid.ci95.0\endcsname{0.000}
\expandafter\def\csname cnt@rev.desc.d6.sspro.p1.C.box.dual-d2.boot-resid.ci95.0\endcsname{0}
\expandafter\def\csname pct@rev.desc.d6.sspro.p1.C.box.dual-d2.boot-resid.ci95.0\endcsname{0.0}
\expandafter\def\csname ppt@rev.desc.d6.sspro.p1.C.box.dual-d2.boot-resid.ci95.0\endcsname{0.0}
\expandafter\def\csname val@rev.desc.d6.sspro.p1.C.box.dual-d2.boot-resid.ci95.1\endcsname{0.000}
\expandafter\def\csname cnt@rev.desc.d6.sspro.p1.C.box.dual-d2.boot-resid.ci95.1\endcsname{0}
\expandafter\def\csname pct@rev.desc.d6.sspro.p1.C.box.dual-d2.boot-resid.ci95.1\endcsname{0.0}
\expandafter\def\csname ppt@rev.desc.d6.sspro.p1.C.box.dual-d2.boot-resid.ci95.1\endcsname{0.0}
\expandafter\def\csname val@rev.desc.d6.sspro.p1.C.box.dual-d2.boot-resid.n-apps\endcsname{6}
\expandafter\def\csname cnt@rev.desc.d6.sspro.p1.C.box.dual-d2.boot-resid.n-apps\endcsname{6}
\expandafter\def\csname pct@rev.desc.d6.sspro.p1.C.box.dual-d2.boot-resid.n-apps\endcsname{600.0}
\expandafter\def\csname ppt@rev.desc.d6.sspro.p1.C.box.dual-d2.boot-resid.n-apps\endcsname{$+$600.0}
\expandafter\def\csname val@rev.desc.d6.sspro.p1.C.box.dual-d2.boot-resid.n-boot-valid\endcsname{4000}
\expandafter\def\csname cnt@rev.desc.d6.sspro.p1.C.box.dual-d2.boot-resid.n-boot-valid\endcsname{4{,}000}
\expandafter\def\csname pct@rev.desc.d6.sspro.p1.C.box.dual-d2.boot-resid.n-boot-valid\endcsname{400000.0}
\expandafter\def\csname ppt@rev.desc.d6.sspro.p1.C.box.dual-d2.boot-resid.n-boot-valid\endcsname{$+$400000.0}
\expandafter\def\csname val@rev.desc.d6.sspro.p1.C.box.dual-d2.boot-resid.value\endcsname{0.000}
\expandafter\def\csname cnt@rev.desc.d6.sspro.p1.C.box.dual-d2.boot-resid.value\endcsname{0}
\expandafter\def\csname pct@rev.desc.d6.sspro.p1.C.box.dual-d2.boot-resid.value\endcsname{0.0}
\expandafter\def\csname ppt@rev.desc.d6.sspro.p1.C.box.dual-d2.boot-resid.value\endcsname{0.0}
\expandafter\def\csname val@rev.desc.d6.sspro.p1.C.box.dual-d2.boot-share.ci95.0\endcsname{0.292}
\expandafter\def\csname cnt@rev.desc.d6.sspro.p1.C.box.dual-d2.boot-share.ci95.0\endcsname{0}
\expandafter\def\csname pct@rev.desc.d6.sspro.p1.C.box.dual-d2.boot-share.ci95.0\endcsname{29.2}
\expandafter\def\csname ppt@rev.desc.d6.sspro.p1.C.box.dual-d2.boot-share.ci95.0\endcsname{$+$29.2}
\expandafter\def\csname val@rev.desc.d6.sspro.p1.C.box.dual-d2.boot-share.ci95.1\endcsname{0.729}
\expandafter\def\csname cnt@rev.desc.d6.sspro.p1.C.box.dual-d2.boot-share.ci95.1\endcsname{1}
\expandafter\def\csname pct@rev.desc.d6.sspro.p1.C.box.dual-d2.boot-share.ci95.1\endcsname{72.9}
\expandafter\def\csname ppt@rev.desc.d6.sspro.p1.C.box.dual-d2.boot-share.ci95.1\endcsname{$+$72.9}
\expandafter\def\csname val@rev.desc.d6.sspro.p1.C.box.dual-d2.boot-share.n-apps\endcsname{6}
\expandafter\def\csname cnt@rev.desc.d6.sspro.p1.C.box.dual-d2.boot-share.n-apps\endcsname{6}
\expandafter\def\csname pct@rev.desc.d6.sspro.p1.C.box.dual-d2.boot-share.n-apps\endcsname{600.0}
\expandafter\def\csname ppt@rev.desc.d6.sspro.p1.C.box.dual-d2.boot-share.n-apps\endcsname{$+$600.0}
\expandafter\def\csname val@rev.desc.d6.sspro.p1.C.box.dual-d2.boot-share.n-boot-valid\endcsname{4000}
\expandafter\def\csname cnt@rev.desc.d6.sspro.p1.C.box.dual-d2.boot-share.n-boot-valid\endcsname{4{,}000}
\expandafter\def\csname pct@rev.desc.d6.sspro.p1.C.box.dual-d2.boot-share.n-boot-valid\endcsname{400000.0}
\expandafter\def\csname ppt@rev.desc.d6.sspro.p1.C.box.dual-d2.boot-share.n-boot-valid\endcsname{$+$400000.0}
\expandafter\def\csname val@rev.desc.d6.sspro.p1.C.box.dual-d2.boot-share.value\endcsname{0.521}
\expandafter\def\csname cnt@rev.desc.d6.sspro.p1.C.box.dual-d2.boot-share.value\endcsname{1}
\expandafter\def\csname pct@rev.desc.d6.sspro.p1.C.box.dual-d2.boot-share.value\endcsname{52.1}
\expandafter\def\csname ppt@rev.desc.d6.sspro.p1.C.box.dual-d2.boot-share.value\endcsname{$+$52.1}
\expandafter\def\csname val@rev.desc.d6.sspro.p1.C.box.dual-d2.n\endcsname{48}
\expandafter\def\csname cnt@rev.desc.d6.sspro.p1.C.box.dual-d2.n\endcsname{48}
\expandafter\def\csname pct@rev.desc.d6.sspro.p1.C.box.dual-d2.n\endcsname{4800.0}
\expandafter\def\csname ppt@rev.desc.d6.sspro.p1.C.box.dual-d2.n\endcsname{$+$4800.0}
\expandafter\def\csname val@rev.desc.d6.sspro.p1.C.box.dual-d2.resid-agree-stable\endcsname{0.000}
\expandafter\def\csname cnt@rev.desc.d6.sspro.p1.C.box.dual-d2.resid-agree-stable\endcsname{0}
\expandafter\def\csname pct@rev.desc.d6.sspro.p1.C.box.dual-d2.resid-agree-stable\endcsname{0.0}
\expandafter\def\csname ppt@rev.desc.d6.sspro.p1.C.box.dual-d2.resid-agree-stable\endcsname{0.0}
\expandafter\def\csname val@rev.desc.d6.sspro.p1.C.box.dual-d2.resid-disagree\endcsname{0.150}
\expandafter\def\csname cnt@rev.desc.d6.sspro.p1.C.box.dual-d2.resid-disagree\endcsname{0}
\expandafter\def\csname pct@rev.desc.d6.sspro.p1.C.box.dual-d2.resid-disagree\endcsname{15.0}
\expandafter\def\csname ppt@rev.desc.d6.sspro.p1.C.box.dual-d2.resid-disagree\endcsname{$+$15.0}
\expandafter\def\csname val@rev.desc.d6.sspro.p1.C.box.dual-d2.share-agree-stable\endcsname{0.521}
\expandafter\def\csname cnt@rev.desc.d6.sspro.p1.C.box.dual-d2.share-agree-stable\endcsname{1}
\expandafter\def\csname pct@rev.desc.d6.sspro.p1.C.box.dual-d2.share-agree-stable\endcsname{52.1}
\expandafter\def\csname ppt@rev.desc.d6.sspro.p1.C.box.dual-d2.share-agree-stable\endcsname{$+$52.1}
\expandafter\def\csname val@rev.desc.d6.sspro.p1.C.box.grid2-d3.N-agree-stable\endcsname{28}
\expandafter\def\csname cnt@rev.desc.d6.sspro.p1.C.box.grid2-d3.N-agree-stable\endcsname{28}
\expandafter\def\csname pct@rev.desc.d6.sspro.p1.C.box.grid2-d3.N-agree-stable\endcsname{2800.0}
\expandafter\def\csname ppt@rev.desc.d6.sspro.p1.C.box.grid2-d3.N-agree-stable\endcsname{$+$2800.0}
\expandafter\def\csname val@rev.desc.d6.sspro.p1.C.box.grid2-d3.N-disagree\endcsname{18}
\expandafter\def\csname cnt@rev.desc.d6.sspro.p1.C.box.grid2-d3.N-disagree\endcsname{18}
\expandafter\def\csname pct@rev.desc.d6.sspro.p1.C.box.grid2-d3.N-disagree\endcsname{1800.0}
\expandafter\def\csname ppt@rev.desc.d6.sspro.p1.C.box.grid2-d3.N-disagree\endcsname{$+$1800.0}
\expandafter\def\csname val@rev.desc.d6.sspro.p1.C.box.grid2-d3.boot-resid.ci95.0\endcsname{-0.111}
\expandafter\def\csname cnt@rev.desc.d6.sspro.p1.C.box.grid2-d3.boot-resid.ci95.0\endcsname{0}
\expandafter\def\csname pct@rev.desc.d6.sspro.p1.C.box.grid2-d3.boot-resid.ci95.0\endcsname{$-$11.1}
\expandafter\def\csname ppt@rev.desc.d6.sspro.p1.C.box.grid2-d3.boot-resid.ci95.0\endcsname{$-$11.1}
\expandafter\def\csname val@rev.desc.d6.sspro.p1.C.box.grid2-d3.boot-resid.ci95.1\endcsname{0.000}
\expandafter\def\csname cnt@rev.desc.d6.sspro.p1.C.box.grid2-d3.boot-resid.ci95.1\endcsname{0}
\expandafter\def\csname pct@rev.desc.d6.sspro.p1.C.box.grid2-d3.boot-resid.ci95.1\endcsname{0.0}
\expandafter\def\csname ppt@rev.desc.d6.sspro.p1.C.box.grid2-d3.boot-resid.ci95.1\endcsname{0.0}
\expandafter\def\csname val@rev.desc.d6.sspro.p1.C.box.grid2-d3.boot-resid.n-apps\endcsname{6}
\expandafter\def\csname cnt@rev.desc.d6.sspro.p1.C.box.grid2-d3.boot-resid.n-apps\endcsname{6}
\expandafter\def\csname pct@rev.desc.d6.sspro.p1.C.box.grid2-d3.boot-resid.n-apps\endcsname{600.0}
\expandafter\def\csname ppt@rev.desc.d6.sspro.p1.C.box.grid2-d3.boot-resid.n-apps\endcsname{$+$600.0}
\expandafter\def\csname val@rev.desc.d6.sspro.p1.C.box.grid2-d3.boot-resid.n-boot-valid\endcsname{4000}
\expandafter\def\csname cnt@rev.desc.d6.sspro.p1.C.box.grid2-d3.boot-resid.n-boot-valid\endcsname{4{,}000}
\expandafter\def\csname pct@rev.desc.d6.sspro.p1.C.box.grid2-d3.boot-resid.n-boot-valid\endcsname{400000.0}
\expandafter\def\csname ppt@rev.desc.d6.sspro.p1.C.box.grid2-d3.boot-resid.n-boot-valid\endcsname{$+$400000.0}
\expandafter\def\csname val@rev.desc.d6.sspro.p1.C.box.grid2-d3.boot-resid.value\endcsname{-0.036}
\expandafter\def\csname cnt@rev.desc.d6.sspro.p1.C.box.grid2-d3.boot-resid.value\endcsname{0}
\expandafter\def\csname pct@rev.desc.d6.sspro.p1.C.box.grid2-d3.boot-resid.value\endcsname{$-$3.6}
\expandafter\def\csname ppt@rev.desc.d6.sspro.p1.C.box.grid2-d3.boot-resid.value\endcsname{$-$3.6}
\expandafter\def\csname val@rev.desc.d6.sspro.p1.C.box.grid2-d3.boot-share.ci95.0\endcsname{0.375}
\expandafter\def\csname cnt@rev.desc.d6.sspro.p1.C.box.grid2-d3.boot-share.ci95.0\endcsname{0}
\expandafter\def\csname pct@rev.desc.d6.sspro.p1.C.box.grid2-d3.boot-share.ci95.0\endcsname{37.5}
\expandafter\def\csname ppt@rev.desc.d6.sspro.p1.C.box.grid2-d3.boot-share.ci95.0\endcsname{$+$37.5}
\expandafter\def\csname val@rev.desc.d6.sspro.p1.C.box.grid2-d3.boot-share.ci95.1\endcsname{0.750}
\expandafter\def\csname cnt@rev.desc.d6.sspro.p1.C.box.grid2-d3.boot-share.ci95.1\endcsname{1}
\expandafter\def\csname pct@rev.desc.d6.sspro.p1.C.box.grid2-d3.boot-share.ci95.1\endcsname{75.0}
\expandafter\def\csname ppt@rev.desc.d6.sspro.p1.C.box.grid2-d3.boot-share.ci95.1\endcsname{$+$75.0}
\expandafter\def\csname val@rev.desc.d6.sspro.p1.C.box.grid2-d3.boot-share.n-apps\endcsname{6}
\expandafter\def\csname cnt@rev.desc.d6.sspro.p1.C.box.grid2-d3.boot-share.n-apps\endcsname{6}
\expandafter\def\csname pct@rev.desc.d6.sspro.p1.C.box.grid2-d3.boot-share.n-apps\endcsname{600.0}
\expandafter\def\csname ppt@rev.desc.d6.sspro.p1.C.box.grid2-d3.boot-share.n-apps\endcsname{$+$600.0}
\expandafter\def\csname val@rev.desc.d6.sspro.p1.C.box.grid2-d3.boot-share.n-boot-valid\endcsname{4000}
\expandafter\def\csname cnt@rev.desc.d6.sspro.p1.C.box.grid2-d3.boot-share.n-boot-valid\endcsname{4{,}000}
\expandafter\def\csname pct@rev.desc.d6.sspro.p1.C.box.grid2-d3.boot-share.n-boot-valid\endcsname{400000.0}
\expandafter\def\csname ppt@rev.desc.d6.sspro.p1.C.box.grid2-d3.boot-share.n-boot-valid\endcsname{$+$400000.0}
\expandafter\def\csname val@rev.desc.d6.sspro.p1.C.box.grid2-d3.boot-share.value\endcsname{0.583}
\expandafter\def\csname cnt@rev.desc.d6.sspro.p1.C.box.grid2-d3.boot-share.value\endcsname{1}
\expandafter\def\csname pct@rev.desc.d6.sspro.p1.C.box.grid2-d3.boot-share.value\endcsname{58.3}
\expandafter\def\csname ppt@rev.desc.d6.sspro.p1.C.box.grid2-d3.boot-share.value\endcsname{$+$58.3}
\expandafter\def\csname val@rev.desc.d6.sspro.p1.C.box.grid2-d3.n\endcsname{48}
\expandafter\def\csname cnt@rev.desc.d6.sspro.p1.C.box.grid2-d3.n\endcsname{48}
\expandafter\def\csname pct@rev.desc.d6.sspro.p1.C.box.grid2-d3.n\endcsname{4800.0}
\expandafter\def\csname ppt@rev.desc.d6.sspro.p1.C.box.grid2-d3.n\endcsname{$+$4800.0}
\expandafter\def\csname val@rev.desc.d6.sspro.p1.C.box.grid2-d3.resid-agree-stable\endcsname{-0.036}
\expandafter\def\csname cnt@rev.desc.d6.sspro.p1.C.box.grid2-d3.resid-agree-stable\endcsname{0}
\expandafter\def\csname pct@rev.desc.d6.sspro.p1.C.box.grid2-d3.resid-agree-stable\endcsname{$-$3.6}
\expandafter\def\csname ppt@rev.desc.d6.sspro.p1.C.box.grid2-d3.resid-agree-stable\endcsname{$-$3.6}
\expandafter\def\csname val@rev.desc.d6.sspro.p1.C.box.grid2-d3.resid-disagree\endcsname{0.056}
\expandafter\def\csname cnt@rev.desc.d6.sspro.p1.C.box.grid2-d3.resid-disagree\endcsname{0}
\expandafter\def\csname pct@rev.desc.d6.sspro.p1.C.box.grid2-d3.resid-disagree\endcsname{5.6}
\expandafter\def\csname ppt@rev.desc.d6.sspro.p1.C.box.grid2-d3.resid-disagree\endcsname{$+$5.6}
\expandafter\def\csname val@rev.desc.d6.sspro.p1.C.box.grid2-d3.share-agree-stable\endcsname{0.583}
\expandafter\def\csname cnt@rev.desc.d6.sspro.p1.C.box.grid2-d3.share-agree-stable\endcsname{1}
\expandafter\def\csname pct@rev.desc.d6.sspro.p1.C.box.grid2-d3.share-agree-stable\endcsname{58.3}
\expandafter\def\csname ppt@rev.desc.d6.sspro.p1.C.box.grid2-d3.share-agree-stable\endcsname{$+$58.3}
\expandafter\def\csname val@rev.desc.d6.sspro.p1.C.box.zoomclick-d3.N-agree-stable\endcsname{30}
\expandafter\def\csname cnt@rev.desc.d6.sspro.p1.C.box.zoomclick-d3.N-agree-stable\endcsname{30}
\expandafter\def\csname pct@rev.desc.d6.sspro.p1.C.box.zoomclick-d3.N-agree-stable\endcsname{3000.0}
\expandafter\def\csname ppt@rev.desc.d6.sspro.p1.C.box.zoomclick-d3.N-agree-stable\endcsname{$+$3000.0}
\expandafter\def\csname val@rev.desc.d6.sspro.p1.C.box.zoomclick-d3.N-disagree\endcsname{17}
\expandafter\def\csname cnt@rev.desc.d6.sspro.p1.C.box.zoomclick-d3.N-disagree\endcsname{17}
\expandafter\def\csname pct@rev.desc.d6.sspro.p1.C.box.zoomclick-d3.N-disagree\endcsname{1700.0}
\expandafter\def\csname ppt@rev.desc.d6.sspro.p1.C.box.zoomclick-d3.N-disagree\endcsname{$+$1700.0}
\expandafter\def\csname val@rev.desc.d6.sspro.p1.C.box.zoomclick-d3.boot-resid.ci95.0\endcsname{0.000}
\expandafter\def\csname cnt@rev.desc.d6.sspro.p1.C.box.zoomclick-d3.boot-resid.ci95.0\endcsname{0}
\expandafter\def\csname pct@rev.desc.d6.sspro.p1.C.box.zoomclick-d3.boot-resid.ci95.0\endcsname{0.0}
\expandafter\def\csname ppt@rev.desc.d6.sspro.p1.C.box.zoomclick-d3.boot-resid.ci95.0\endcsname{0.0}
\expandafter\def\csname val@rev.desc.d6.sspro.p1.C.box.zoomclick-d3.boot-resid.ci95.1\endcsname{0.000}
\expandafter\def\csname cnt@rev.desc.d6.sspro.p1.C.box.zoomclick-d3.boot-resid.ci95.1\endcsname{0}
\expandafter\def\csname pct@rev.desc.d6.sspro.p1.C.box.zoomclick-d3.boot-resid.ci95.1\endcsname{0.0}
\expandafter\def\csname ppt@rev.desc.d6.sspro.p1.C.box.zoomclick-d3.boot-resid.ci95.1\endcsname{0.0}
\expandafter\def\csname val@rev.desc.d6.sspro.p1.C.box.zoomclick-d3.boot-resid.n-apps\endcsname{6}
\expandafter\def\csname cnt@rev.desc.d6.sspro.p1.C.box.zoomclick-d3.boot-resid.n-apps\endcsname{6}
\expandafter\def\csname pct@rev.desc.d6.sspro.p1.C.box.zoomclick-d3.boot-resid.n-apps\endcsname{600.0}
\expandafter\def\csname ppt@rev.desc.d6.sspro.p1.C.box.zoomclick-d3.boot-resid.n-apps\endcsname{$+$600.0}
\expandafter\def\csname val@rev.desc.d6.sspro.p1.C.box.zoomclick-d3.boot-resid.n-boot-valid\endcsname{4000}
\expandafter\def\csname cnt@rev.desc.d6.sspro.p1.C.box.zoomclick-d3.boot-resid.n-boot-valid\endcsname{4{,}000}
\expandafter\def\csname pct@rev.desc.d6.sspro.p1.C.box.zoomclick-d3.boot-resid.n-boot-valid\endcsname{400000.0}
\expandafter\def\csname ppt@rev.desc.d6.sspro.p1.C.box.zoomclick-d3.boot-resid.n-boot-valid\endcsname{$+$400000.0}
\expandafter\def\csname val@rev.desc.d6.sspro.p1.C.box.zoomclick-d3.boot-resid.value\endcsname{0.000}
\expandafter\def\csname cnt@rev.desc.d6.sspro.p1.C.box.zoomclick-d3.boot-resid.value\endcsname{0}
\expandafter\def\csname pct@rev.desc.d6.sspro.p1.C.box.zoomclick-d3.boot-resid.value\endcsname{0.0}
\expandafter\def\csname ppt@rev.desc.d6.sspro.p1.C.box.zoomclick-d3.boot-resid.value\endcsname{0.0}
\expandafter\def\csname val@rev.desc.d6.sspro.p1.C.box.zoomclick-d3.boot-share.ci95.0\endcsname{0.396}
\expandafter\def\csname cnt@rev.desc.d6.sspro.p1.C.box.zoomclick-d3.boot-share.ci95.0\endcsname{0}
\expandafter\def\csname pct@rev.desc.d6.sspro.p1.C.box.zoomclick-d3.boot-share.ci95.0\endcsname{39.6}
\expandafter\def\csname ppt@rev.desc.d6.sspro.p1.C.box.zoomclick-d3.boot-share.ci95.0\endcsname{$+$39.6}
\expandafter\def\csname val@rev.desc.d6.sspro.p1.C.box.zoomclick-d3.boot-share.ci95.1\endcsname{0.812}
\expandafter\def\csname cnt@rev.desc.d6.sspro.p1.C.box.zoomclick-d3.boot-share.ci95.1\endcsname{1}
\expandafter\def\csname pct@rev.desc.d6.sspro.p1.C.box.zoomclick-d3.boot-share.ci95.1\endcsname{81.2}
\expandafter\def\csname ppt@rev.desc.d6.sspro.p1.C.box.zoomclick-d3.boot-share.ci95.1\endcsname{$+$81.2}
\expandafter\def\csname val@rev.desc.d6.sspro.p1.C.box.zoomclick-d3.boot-share.n-apps\endcsname{6}
\expandafter\def\csname cnt@rev.desc.d6.sspro.p1.C.box.zoomclick-d3.boot-share.n-apps\endcsname{6}
\expandafter\def\csname pct@rev.desc.d6.sspro.p1.C.box.zoomclick-d3.boot-share.n-apps\endcsname{600.0}
\expandafter\def\csname ppt@rev.desc.d6.sspro.p1.C.box.zoomclick-d3.boot-share.n-apps\endcsname{$+$600.0}
\expandafter\def\csname val@rev.desc.d6.sspro.p1.C.box.zoomclick-d3.boot-share.n-boot-valid\endcsname{4000}
\expandafter\def\csname cnt@rev.desc.d6.sspro.p1.C.box.zoomclick-d3.boot-share.n-boot-valid\endcsname{4{,}000}
\expandafter\def\csname pct@rev.desc.d6.sspro.p1.C.box.zoomclick-d3.boot-share.n-boot-valid\endcsname{400000.0}
\expandafter\def\csname ppt@rev.desc.d6.sspro.p1.C.box.zoomclick-d3.boot-share.n-boot-valid\endcsname{$+$400000.0}
\expandafter\def\csname val@rev.desc.d6.sspro.p1.C.box.zoomclick-d3.boot-share.value\endcsname{0.625}
\expandafter\def\csname cnt@rev.desc.d6.sspro.p1.C.box.zoomclick-d3.boot-share.value\endcsname{1}
\expandafter\def\csname pct@rev.desc.d6.sspro.p1.C.box.zoomclick-d3.boot-share.value\endcsname{62.5}
\expandafter\def\csname ppt@rev.desc.d6.sspro.p1.C.box.zoomclick-d3.boot-share.value\endcsname{$+$62.5}
\expandafter\def\csname val@rev.desc.d6.sspro.p1.C.box.zoomclick-d3.n\endcsname{48}
\expandafter\def\csname cnt@rev.desc.d6.sspro.p1.C.box.zoomclick-d3.n\endcsname{48}
\expandafter\def\csname pct@rev.desc.d6.sspro.p1.C.box.zoomclick-d3.n\endcsname{4800.0}
\expandafter\def\csname ppt@rev.desc.d6.sspro.p1.C.box.zoomclick-d3.n\endcsname{$+$4800.0}
\expandafter\def\csname val@rev.desc.d6.sspro.p1.C.box.zoomclick-d3.resid-agree-stable\endcsname{0.000}
\expandafter\def\csname cnt@rev.desc.d6.sspro.p1.C.box.zoomclick-d3.resid-agree-stable\endcsname{0}
\expandafter\def\csname pct@rev.desc.d6.sspro.p1.C.box.zoomclick-d3.resid-agree-stable\endcsname{0.0}
\expandafter\def\csname ppt@rev.desc.d6.sspro.p1.C.box.zoomclick-d3.resid-agree-stable\endcsname{0.0}
\expandafter\def\csname val@rev.desc.d6.sspro.p1.C.box.zoomclick-d3.resid-disagree\endcsname{0.059}
\expandafter\def\csname cnt@rev.desc.d6.sspro.p1.C.box.zoomclick-d3.resid-disagree\endcsname{0}
\expandafter\def\csname pct@rev.desc.d6.sspro.p1.C.box.zoomclick-d3.resid-disagree\endcsname{5.9}
\expandafter\def\csname ppt@rev.desc.d6.sspro.p1.C.box.zoomclick-d3.resid-disagree\endcsname{$+$5.9}
\expandafter\def\csname val@rev.desc.d6.sspro.p1.C.box.zoomclick-d3.share-agree-stable\endcsname{0.625}
\expandafter\def\csname cnt@rev.desc.d6.sspro.p1.C.box.zoomclick-d3.share-agree-stable\endcsname{1}
\expandafter\def\csname pct@rev.desc.d6.sspro.p1.C.box.zoomclick-d3.share-agree-stable\endcsname{62.5}
\expandafter\def\csname ppt@rev.desc.d6.sspro.p1.C.box.zoomclick-d3.share-agree-stable\endcsname{$+$62.5}
\expandafter\def\csname val@rev.desc.d6.sspro.p1.C.tdiag.catalog-pooled.N-agree-stable\endcsname{143}
\expandafter\def\csname cnt@rev.desc.d6.sspro.p1.C.tdiag.catalog-pooled.N-agree-stable\endcsname{143}
\expandafter\def\csname pct@rev.desc.d6.sspro.p1.C.tdiag.catalog-pooled.N-agree-stable\endcsname{14300.0}
\expandafter\def\csname ppt@rev.desc.d6.sspro.p1.C.tdiag.catalog-pooled.N-agree-stable\endcsname{$+$14300.0}
\expandafter\def\csname val@rev.desc.d6.sspro.p1.C.tdiag.catalog-pooled.N-disagree\endcsname{39}
\expandafter\def\csname cnt@rev.desc.d6.sspro.p1.C.tdiag.catalog-pooled.N-disagree\endcsname{39}
\expandafter\def\csname pct@rev.desc.d6.sspro.p1.C.tdiag.catalog-pooled.N-disagree\endcsname{3900.0}
\expandafter\def\csname ppt@rev.desc.d6.sspro.p1.C.tdiag.catalog-pooled.N-disagree\endcsname{$+$3900.0}
\expandafter\def\csname val@rev.desc.d6.sspro.p1.C.tdiag.catalog-pooled.boot-resid.ci95.0\endcsname{-0.006}
\expandafter\def\csname cnt@rev.desc.d6.sspro.p1.C.tdiag.catalog-pooled.boot-resid.ci95.0\endcsname{0}
\expandafter\def\csname pct@rev.desc.d6.sspro.p1.C.tdiag.catalog-pooled.boot-resid.ci95.0\endcsname{$-$0.6}
\expandafter\def\csname ppt@rev.desc.d6.sspro.p1.C.tdiag.catalog-pooled.boot-resid.ci95.0\endcsname{$-$0.6}
\expandafter\def\csname val@rev.desc.d6.sspro.p1.C.tdiag.catalog-pooled.boot-resid.ci95.1\endcsname{0.061}
\expandafter\def\csname cnt@rev.desc.d6.sspro.p1.C.tdiag.catalog-pooled.boot-resid.ci95.1\endcsname{0}
\expandafter\def\csname pct@rev.desc.d6.sspro.p1.C.tdiag.catalog-pooled.boot-resid.ci95.1\endcsname{6.1}
\expandafter\def\csname ppt@rev.desc.d6.sspro.p1.C.tdiag.catalog-pooled.boot-resid.ci95.1\endcsname{$+$6.1}
\expandafter\def\csname val@rev.desc.d6.sspro.p1.C.tdiag.catalog-pooled.boot-resid.n-apps\endcsname{6}
\expandafter\def\csname cnt@rev.desc.d6.sspro.p1.C.tdiag.catalog-pooled.boot-resid.n-apps\endcsname{6}
\expandafter\def\csname pct@rev.desc.d6.sspro.p1.C.tdiag.catalog-pooled.boot-resid.n-apps\endcsname{600.0}
\expandafter\def\csname ppt@rev.desc.d6.sspro.p1.C.tdiag.catalog-pooled.boot-resid.n-apps\endcsname{$+$600.0}
\expandafter\def\csname val@rev.desc.d6.sspro.p1.C.tdiag.catalog-pooled.boot-resid.n-boot-valid\endcsname{4000}
\expandafter\def\csname cnt@rev.desc.d6.sspro.p1.C.tdiag.catalog-pooled.boot-resid.n-boot-valid\endcsname{4{,}000}
\expandafter\def\csname pct@rev.desc.d6.sspro.p1.C.tdiag.catalog-pooled.boot-resid.n-boot-valid\endcsname{400000.0}
\expandafter\def\csname ppt@rev.desc.d6.sspro.p1.C.tdiag.catalog-pooled.boot-resid.n-boot-valid\endcsname{$+$400000.0}
\expandafter\def\csname val@rev.desc.d6.sspro.p1.C.tdiag.catalog-pooled.boot-resid.value\endcsname{0.028}
\expandafter\def\csname cnt@rev.desc.d6.sspro.p1.C.tdiag.catalog-pooled.boot-resid.value\endcsname{0}
\expandafter\def\csname pct@rev.desc.d6.sspro.p1.C.tdiag.catalog-pooled.boot-resid.value\endcsname{2.8}
\expandafter\def\csname ppt@rev.desc.d6.sspro.p1.C.tdiag.catalog-pooled.boot-resid.value\endcsname{$+$2.8}
\expandafter\def\csname val@rev.desc.d6.sspro.p1.C.tdiag.catalog-pooled.boot-share.ci95.0\endcsname{0.547}
\expandafter\def\csname cnt@rev.desc.d6.sspro.p1.C.tdiag.catalog-pooled.boot-share.ci95.0\endcsname{1}
\expandafter\def\csname pct@rev.desc.d6.sspro.p1.C.tdiag.catalog-pooled.boot-share.ci95.0\endcsname{54.7}
\expandafter\def\csname ppt@rev.desc.d6.sspro.p1.C.tdiag.catalog-pooled.boot-share.ci95.0\endcsname{$+$54.7}
\expandafter\def\csname val@rev.desc.d6.sspro.p1.C.tdiag.catalog-pooled.boot-share.ci95.1\endcsname{0.906}
\expandafter\def\csname cnt@rev.desc.d6.sspro.p1.C.tdiag.catalog-pooled.boot-share.ci95.1\endcsname{1}
\expandafter\def\csname pct@rev.desc.d6.sspro.p1.C.tdiag.catalog-pooled.boot-share.ci95.1\endcsname{90.6}
\expandafter\def\csname ppt@rev.desc.d6.sspro.p1.C.tdiag.catalog-pooled.boot-share.ci95.1\endcsname{$+$90.6}
\expandafter\def\csname val@rev.desc.d6.sspro.p1.C.tdiag.catalog-pooled.boot-share.n-apps\endcsname{6}
\expandafter\def\csname cnt@rev.desc.d6.sspro.p1.C.tdiag.catalog-pooled.boot-share.n-apps\endcsname{6}
\expandafter\def\csname pct@rev.desc.d6.sspro.p1.C.tdiag.catalog-pooled.boot-share.n-apps\endcsname{600.0}
\expandafter\def\csname ppt@rev.desc.d6.sspro.p1.C.tdiag.catalog-pooled.boot-share.n-apps\endcsname{$+$600.0}
\expandafter\def\csname val@rev.desc.d6.sspro.p1.C.tdiag.catalog-pooled.boot-share.n-boot-valid\endcsname{4000}
\expandafter\def\csname cnt@rev.desc.d6.sspro.p1.C.tdiag.catalog-pooled.boot-share.n-boot-valid\endcsname{4{,}000}
\expandafter\def\csname pct@rev.desc.d6.sspro.p1.C.tdiag.catalog-pooled.boot-share.n-boot-valid\endcsname{400000.0}
\expandafter\def\csname ppt@rev.desc.d6.sspro.p1.C.tdiag.catalog-pooled.boot-share.n-boot-valid\endcsname{$+$400000.0}
\expandafter\def\csname val@rev.desc.d6.sspro.p1.C.tdiag.catalog-pooled.boot-share.value\endcsname{0.745}
\expandafter\def\csname cnt@rev.desc.d6.sspro.p1.C.tdiag.catalog-pooled.boot-share.value\endcsname{1}
\expandafter\def\csname pct@rev.desc.d6.sspro.p1.C.tdiag.catalog-pooled.boot-share.value\endcsname{74.5}
\expandafter\def\csname ppt@rev.desc.d6.sspro.p1.C.tdiag.catalog-pooled.boot-share.value\endcsname{$+$74.5}
\expandafter\def\csname val@rev.desc.d6.sspro.p1.C.tdiag.catalog-pooled.gap\endcsname{-0.028}
\expandafter\def\csname cnt@rev.desc.d6.sspro.p1.C.tdiag.catalog-pooled.gap\endcsname{0}
\expandafter\def\csname pct@rev.desc.d6.sspro.p1.C.tdiag.catalog-pooled.gap\endcsname{$-$2.8}
\expandafter\def\csname ppt@rev.desc.d6.sspro.p1.C.tdiag.catalog-pooled.gap\endcsname{$-$2.8}
\expandafter\def\csname val@rev.desc.d6.sspro.p1.C.tdiag.catalog-pooled.n\endcsname{192}
\expandafter\def\csname cnt@rev.desc.d6.sspro.p1.C.tdiag.catalog-pooled.n\endcsname{192}
\expandafter\def\csname pct@rev.desc.d6.sspro.p1.C.tdiag.catalog-pooled.n\endcsname{19200.0}
\expandafter\def\csname ppt@rev.desc.d6.sspro.p1.C.tdiag.catalog-pooled.n\endcsname{$+$19200.0}
\expandafter\def\csname val@rev.desc.d6.sspro.p1.C.tdiag.catalog-pooled.resid-agree-stable\endcsname{0.028}
\expandafter\def\csname cnt@rev.desc.d6.sspro.p1.C.tdiag.catalog-pooled.resid-agree-stable\endcsname{0}
\expandafter\def\csname pct@rev.desc.d6.sspro.p1.C.tdiag.catalog-pooled.resid-agree-stable\endcsname{2.8}
\expandafter\def\csname ppt@rev.desc.d6.sspro.p1.C.tdiag.catalog-pooled.resid-agree-stable\endcsname{$+$2.8}
\expandafter\def\csname val@rev.desc.d6.sspro.p1.C.tdiag.catalog-pooled.resid-disagree\endcsname{0.000}
\expandafter\def\csname cnt@rev.desc.d6.sspro.p1.C.tdiag.catalog-pooled.resid-disagree\endcsname{0}
\expandafter\def\csname pct@rev.desc.d6.sspro.p1.C.tdiag.catalog-pooled.resid-disagree\endcsname{0.0}
\expandafter\def\csname ppt@rev.desc.d6.sspro.p1.C.tdiag.catalog-pooled.resid-disagree\endcsname{0.0}
\expandafter\def\csname val@rev.desc.d6.sspro.p1.C.tdiag.catalog-pooled.share-agree-stable\endcsname{0.745}
\expandafter\def\csname cnt@rev.desc.d6.sspro.p1.C.tdiag.catalog-pooled.share-agree-stable\endcsname{1}
\expandafter\def\csname pct@rev.desc.d6.sspro.p1.C.tdiag.catalog-pooled.share-agree-stable\endcsname{74.5}
\expandafter\def\csname ppt@rev.desc.d6.sspro.p1.C.tdiag.catalog-pooled.share-agree-stable\endcsname{$+$74.5}
\expandafter\def\csname val@rev.desc.d6.sspro.p1.C.tdiag.center-d1.N-agree-stable\endcsname{37}
\expandafter\def\csname cnt@rev.desc.d6.sspro.p1.C.tdiag.center-d1.N-agree-stable\endcsname{37}
\expandafter\def\csname pct@rev.desc.d6.sspro.p1.C.tdiag.center-d1.N-agree-stable\endcsname{3700.0}
\expandafter\def\csname ppt@rev.desc.d6.sspro.p1.C.tdiag.center-d1.N-agree-stable\endcsname{$+$3700.0}
\expandafter\def\csname val@rev.desc.d6.sspro.p1.C.tdiag.center-d1.N-disagree\endcsname{10}
\expandafter\def\csname cnt@rev.desc.d6.sspro.p1.C.tdiag.center-d1.N-disagree\endcsname{10}
\expandafter\def\csname pct@rev.desc.d6.sspro.p1.C.tdiag.center-d1.N-disagree\endcsname{1000.0}
\expandafter\def\csname ppt@rev.desc.d6.sspro.p1.C.tdiag.center-d1.N-disagree\endcsname{$+$1000.0}
\expandafter\def\csname val@rev.desc.d6.sspro.p1.C.tdiag.center-d1.boot-resid.ci95.0\endcsname{0.000}
\expandafter\def\csname cnt@rev.desc.d6.sspro.p1.C.tdiag.center-d1.boot-resid.ci95.0\endcsname{0}
\expandafter\def\csname pct@rev.desc.d6.sspro.p1.C.tdiag.center-d1.boot-resid.ci95.0\endcsname{0.0}
\expandafter\def\csname ppt@rev.desc.d6.sspro.p1.C.tdiag.center-d1.boot-resid.ci95.0\endcsname{0.0}
\expandafter\def\csname val@rev.desc.d6.sspro.p1.C.tdiag.center-d1.boot-resid.ci95.1\endcsname{0.114}
\expandafter\def\csname cnt@rev.desc.d6.sspro.p1.C.tdiag.center-d1.boot-resid.ci95.1\endcsname{0}
\expandafter\def\csname pct@rev.desc.d6.sspro.p1.C.tdiag.center-d1.boot-resid.ci95.1\endcsname{11.4}
\expandafter\def\csname ppt@rev.desc.d6.sspro.p1.C.tdiag.center-d1.boot-resid.ci95.1\endcsname{$+$11.4}
\expandafter\def\csname val@rev.desc.d6.sspro.p1.C.tdiag.center-d1.boot-resid.n-apps\endcsname{6}
\expandafter\def\csname cnt@rev.desc.d6.sspro.p1.C.tdiag.center-d1.boot-resid.n-apps\endcsname{6}
\expandafter\def\csname pct@rev.desc.d6.sspro.p1.C.tdiag.center-d1.boot-resid.n-apps\endcsname{600.0}
\expandafter\def\csname ppt@rev.desc.d6.sspro.p1.C.tdiag.center-d1.boot-resid.n-apps\endcsname{$+$600.0}
\expandafter\def\csname val@rev.desc.d6.sspro.p1.C.tdiag.center-d1.boot-resid.n-boot-valid\endcsname{4000}
\expandafter\def\csname cnt@rev.desc.d6.sspro.p1.C.tdiag.center-d1.boot-resid.n-boot-valid\endcsname{4{,}000}
\expandafter\def\csname pct@rev.desc.d6.sspro.p1.C.tdiag.center-d1.boot-resid.n-boot-valid\endcsname{400000.0}
\expandafter\def\csname ppt@rev.desc.d6.sspro.p1.C.tdiag.center-d1.boot-resid.n-boot-valid\endcsname{$+$400000.0}
\expandafter\def\csname val@rev.desc.d6.sspro.p1.C.tdiag.center-d1.boot-resid.value\endcsname{0.054}
\expandafter\def\csname cnt@rev.desc.d6.sspro.p1.C.tdiag.center-d1.boot-resid.value\endcsname{0}
\expandafter\def\csname pct@rev.desc.d6.sspro.p1.C.tdiag.center-d1.boot-resid.value\endcsname{5.4}
\expandafter\def\csname ppt@rev.desc.d6.sspro.p1.C.tdiag.center-d1.boot-resid.value\endcsname{$+$5.4}
\expandafter\def\csname val@rev.desc.d6.sspro.p1.C.tdiag.center-d1.boot-share.ci95.0\endcsname{0.583}
\expandafter\def\csname cnt@rev.desc.d6.sspro.p1.C.tdiag.center-d1.boot-share.ci95.0\endcsname{1}
\expandafter\def\csname pct@rev.desc.d6.sspro.p1.C.tdiag.center-d1.boot-share.ci95.0\endcsname{58.3}
\expandafter\def\csname ppt@rev.desc.d6.sspro.p1.C.tdiag.center-d1.boot-share.ci95.0\endcsname{$+$58.3}
\expandafter\def\csname val@rev.desc.d6.sspro.p1.C.tdiag.center-d1.boot-share.ci95.1\endcsname{0.938}
\expandafter\def\csname cnt@rev.desc.d6.sspro.p1.C.tdiag.center-d1.boot-share.ci95.1\endcsname{1}
\expandafter\def\csname pct@rev.desc.d6.sspro.p1.C.tdiag.center-d1.boot-share.ci95.1\endcsname{93.8}
\expandafter\def\csname ppt@rev.desc.d6.sspro.p1.C.tdiag.center-d1.boot-share.ci95.1\endcsname{$+$93.8}
\expandafter\def\csname val@rev.desc.d6.sspro.p1.C.tdiag.center-d1.boot-share.n-apps\endcsname{6}
\expandafter\def\csname cnt@rev.desc.d6.sspro.p1.C.tdiag.center-d1.boot-share.n-apps\endcsname{6}
\expandafter\def\csname pct@rev.desc.d6.sspro.p1.C.tdiag.center-d1.boot-share.n-apps\endcsname{600.0}
\expandafter\def\csname ppt@rev.desc.d6.sspro.p1.C.tdiag.center-d1.boot-share.n-apps\endcsname{$+$600.0}
\expandafter\def\csname val@rev.desc.d6.sspro.p1.C.tdiag.center-d1.boot-share.n-boot-valid\endcsname{4000}
\expandafter\def\csname cnt@rev.desc.d6.sspro.p1.C.tdiag.center-d1.boot-share.n-boot-valid\endcsname{4{,}000}
\expandafter\def\csname pct@rev.desc.d6.sspro.p1.C.tdiag.center-d1.boot-share.n-boot-valid\endcsname{400000.0}
\expandafter\def\csname ppt@rev.desc.d6.sspro.p1.C.tdiag.center-d1.boot-share.n-boot-valid\endcsname{$+$400000.0}
\expandafter\def\csname val@rev.desc.d6.sspro.p1.C.tdiag.center-d1.boot-share.value\endcsname{0.771}
\expandafter\def\csname cnt@rev.desc.d6.sspro.p1.C.tdiag.center-d1.boot-share.value\endcsname{1}
\expandafter\def\csname pct@rev.desc.d6.sspro.p1.C.tdiag.center-d1.boot-share.value\endcsname{77.1}
\expandafter\def\csname ppt@rev.desc.d6.sspro.p1.C.tdiag.center-d1.boot-share.value\endcsname{$+$77.1}
\expandafter\def\csname val@rev.desc.d6.sspro.p1.C.tdiag.center-d1.n\endcsname{48}
\expandafter\def\csname cnt@rev.desc.d6.sspro.p1.C.tdiag.center-d1.n\endcsname{48}
\expandafter\def\csname pct@rev.desc.d6.sspro.p1.C.tdiag.center-d1.n\endcsname{4800.0}
\expandafter\def\csname ppt@rev.desc.d6.sspro.p1.C.tdiag.center-d1.n\endcsname{$+$4800.0}
\expandafter\def\csname val@rev.desc.d6.sspro.p1.C.tdiag.center-d1.resid-agree-stable\endcsname{0.054}
\expandafter\def\csname cnt@rev.desc.d6.sspro.p1.C.tdiag.center-d1.resid-agree-stable\endcsname{0}
\expandafter\def\csname pct@rev.desc.d6.sspro.p1.C.tdiag.center-d1.resid-agree-stable\endcsname{5.4}
\expandafter\def\csname ppt@rev.desc.d6.sspro.p1.C.tdiag.center-d1.resid-agree-stable\endcsname{$+$5.4}
\expandafter\def\csname val@rev.desc.d6.sspro.p1.C.tdiag.center-d1.resid-disagree\endcsname{-0.100}
\expandafter\def\csname cnt@rev.desc.d6.sspro.p1.C.tdiag.center-d1.resid-disagree\endcsname{0}
\expandafter\def\csname pct@rev.desc.d6.sspro.p1.C.tdiag.center-d1.resid-disagree\endcsname{$-$10.0}
\expandafter\def\csname ppt@rev.desc.d6.sspro.p1.C.tdiag.center-d1.resid-disagree\endcsname{$-$10.0}
\expandafter\def\csname val@rev.desc.d6.sspro.p1.C.tdiag.center-d1.share-agree-stable\endcsname{0.771}
\expandafter\def\csname cnt@rev.desc.d6.sspro.p1.C.tdiag.center-d1.share-agree-stable\endcsname{1}
\expandafter\def\csname pct@rev.desc.d6.sspro.p1.C.tdiag.center-d1.share-agree-stable\endcsname{77.1}
\expandafter\def\csname ppt@rev.desc.d6.sspro.p1.C.tdiag.center-d1.share-agree-stable\endcsname{$+$77.1}
\expandafter\def\csname val@rev.desc.d6.sspro.p1.C.tdiag.context-d1.N-agree-stable\endcsname{36}
\expandafter\def\csname cnt@rev.desc.d6.sspro.p1.C.tdiag.context-d1.N-agree-stable\endcsname{36}
\expandafter\def\csname pct@rev.desc.d6.sspro.p1.C.tdiag.context-d1.N-agree-stable\endcsname{3600.0}
\expandafter\def\csname ppt@rev.desc.d6.sspro.p1.C.tdiag.context-d1.N-agree-stable\endcsname{$+$3600.0}
\expandafter\def\csname val@rev.desc.d6.sspro.p1.C.tdiag.context-d1.N-disagree\endcsname{8}
\expandafter\def\csname cnt@rev.desc.d6.sspro.p1.C.tdiag.context-d1.N-disagree\endcsname{8}
\expandafter\def\csname pct@rev.desc.d6.sspro.p1.C.tdiag.context-d1.N-disagree\endcsname{800.0}
\expandafter\def\csname ppt@rev.desc.d6.sspro.p1.C.tdiag.context-d1.N-disagree\endcsname{$+$800.0}
\expandafter\def\csname val@rev.desc.d6.sspro.p1.C.tdiag.context-d1.boot-resid.ci95.0\endcsname{0.000}
\expandafter\def\csname cnt@rev.desc.d6.sspro.p1.C.tdiag.context-d1.boot-resid.ci95.0\endcsname{0}
\expandafter\def\csname pct@rev.desc.d6.sspro.p1.C.tdiag.context-d1.boot-resid.ci95.0\endcsname{0.0}
\expandafter\def\csname ppt@rev.desc.d6.sspro.p1.C.tdiag.context-d1.boot-resid.ci95.0\endcsname{0.0}
\expandafter\def\csname val@rev.desc.d6.sspro.p1.C.tdiag.context-d1.boot-resid.ci95.1\endcsname{0.000}
\expandafter\def\csname cnt@rev.desc.d6.sspro.p1.C.tdiag.context-d1.boot-resid.ci95.1\endcsname{0}
\expandafter\def\csname pct@rev.desc.d6.sspro.p1.C.tdiag.context-d1.boot-resid.ci95.1\endcsname{0.0}
\expandafter\def\csname ppt@rev.desc.d6.sspro.p1.C.tdiag.context-d1.boot-resid.ci95.1\endcsname{0.0}
\expandafter\def\csname val@rev.desc.d6.sspro.p1.C.tdiag.context-d1.boot-resid.n-apps\endcsname{6}
\expandafter\def\csname cnt@rev.desc.d6.sspro.p1.C.tdiag.context-d1.boot-resid.n-apps\endcsname{6}
\expandafter\def\csname pct@rev.desc.d6.sspro.p1.C.tdiag.context-d1.boot-resid.n-apps\endcsname{600.0}
\expandafter\def\csname ppt@rev.desc.d6.sspro.p1.C.tdiag.context-d1.boot-resid.n-apps\endcsname{$+$600.0}
\expandafter\def\csname val@rev.desc.d6.sspro.p1.C.tdiag.context-d1.boot-resid.n-boot-valid\endcsname{4000}
\expandafter\def\csname cnt@rev.desc.d6.sspro.p1.C.tdiag.context-d1.boot-resid.n-boot-valid\endcsname{4{,}000}
\expandafter\def\csname pct@rev.desc.d6.sspro.p1.C.tdiag.context-d1.boot-resid.n-boot-valid\endcsname{400000.0}
\expandafter\def\csname ppt@rev.desc.d6.sspro.p1.C.tdiag.context-d1.boot-resid.n-boot-valid\endcsname{$+$400000.0}
\expandafter\def\csname val@rev.desc.d6.sspro.p1.C.tdiag.context-d1.boot-resid.value\endcsname{0.000}
\expandafter\def\csname cnt@rev.desc.d6.sspro.p1.C.tdiag.context-d1.boot-resid.value\endcsname{0}
\expandafter\def\csname pct@rev.desc.d6.sspro.p1.C.tdiag.context-d1.boot-resid.value\endcsname{0.0}
\expandafter\def\csname ppt@rev.desc.d6.sspro.p1.C.tdiag.context-d1.boot-resid.value\endcsname{0.0}
\expandafter\def\csname val@rev.desc.d6.sspro.p1.C.tdiag.context-d1.boot-share.ci95.0\endcsname{0.562}
\expandafter\def\csname cnt@rev.desc.d6.sspro.p1.C.tdiag.context-d1.boot-share.ci95.0\endcsname{1}
\expandafter\def\csname pct@rev.desc.d6.sspro.p1.C.tdiag.context-d1.boot-share.ci95.0\endcsname{56.2}
\expandafter\def\csname ppt@rev.desc.d6.sspro.p1.C.tdiag.context-d1.boot-share.ci95.0\endcsname{$+$56.2}
\expandafter\def\csname val@rev.desc.d6.sspro.p1.C.tdiag.context-d1.boot-share.ci95.1\endcsname{0.917}
\expandafter\def\csname cnt@rev.desc.d6.sspro.p1.C.tdiag.context-d1.boot-share.ci95.1\endcsname{1}
\expandafter\def\csname pct@rev.desc.d6.sspro.p1.C.tdiag.context-d1.boot-share.ci95.1\endcsname{91.7}
\expandafter\def\csname ppt@rev.desc.d6.sspro.p1.C.tdiag.context-d1.boot-share.ci95.1\endcsname{$+$91.7}
\expandafter\def\csname val@rev.desc.d6.sspro.p1.C.tdiag.context-d1.boot-share.n-apps\endcsname{6}
\expandafter\def\csname cnt@rev.desc.d6.sspro.p1.C.tdiag.context-d1.boot-share.n-apps\endcsname{6}
\expandafter\def\csname pct@rev.desc.d6.sspro.p1.C.tdiag.context-d1.boot-share.n-apps\endcsname{600.0}
\expandafter\def\csname ppt@rev.desc.d6.sspro.p1.C.tdiag.context-d1.boot-share.n-apps\endcsname{$+$600.0}
\expandafter\def\csname val@rev.desc.d6.sspro.p1.C.tdiag.context-d1.boot-share.n-boot-valid\endcsname{4000}
\expandafter\def\csname cnt@rev.desc.d6.sspro.p1.C.tdiag.context-d1.boot-share.n-boot-valid\endcsname{4{,}000}
\expandafter\def\csname pct@rev.desc.d6.sspro.p1.C.tdiag.context-d1.boot-share.n-boot-valid\endcsname{400000.0}
\expandafter\def\csname ppt@rev.desc.d6.sspro.p1.C.tdiag.context-d1.boot-share.n-boot-valid\endcsname{$+$400000.0}
\expandafter\def\csname val@rev.desc.d6.sspro.p1.C.tdiag.context-d1.boot-share.value\endcsname{0.750}
\expandafter\def\csname cnt@rev.desc.d6.sspro.p1.C.tdiag.context-d1.boot-share.value\endcsname{1}
\expandafter\def\csname pct@rev.desc.d6.sspro.p1.C.tdiag.context-d1.boot-share.value\endcsname{75.0}
\expandafter\def\csname ppt@rev.desc.d6.sspro.p1.C.tdiag.context-d1.boot-share.value\endcsname{$+$75.0}
\expandafter\def\csname val@rev.desc.d6.sspro.p1.C.tdiag.context-d1.n\endcsname{48}
\expandafter\def\csname cnt@rev.desc.d6.sspro.p1.C.tdiag.context-d1.n\endcsname{48}
\expandafter\def\csname pct@rev.desc.d6.sspro.p1.C.tdiag.context-d1.n\endcsname{4800.0}
\expandafter\def\csname ppt@rev.desc.d6.sspro.p1.C.tdiag.context-d1.n\endcsname{$+$4800.0}
\expandafter\def\csname val@rev.desc.d6.sspro.p1.C.tdiag.context-d1.resid-agree-stable\endcsname{0.000}
\expandafter\def\csname cnt@rev.desc.d6.sspro.p1.C.tdiag.context-d1.resid-agree-stable\endcsname{0}
\expandafter\def\csname pct@rev.desc.d6.sspro.p1.C.tdiag.context-d1.resid-agree-stable\endcsname{0.0}
\expandafter\def\csname ppt@rev.desc.d6.sspro.p1.C.tdiag.context-d1.resid-agree-stable\endcsname{0.0}
\expandafter\def\csname val@rev.desc.d6.sspro.p1.C.tdiag.context-d1.resid-disagree\endcsname{0.000}
\expandafter\def\csname cnt@rev.desc.d6.sspro.p1.C.tdiag.context-d1.resid-disagree\endcsname{0}
\expandafter\def\csname pct@rev.desc.d6.sspro.p1.C.tdiag.context-d1.resid-disagree\endcsname{0.0}
\expandafter\def\csname ppt@rev.desc.d6.sspro.p1.C.tdiag.context-d1.resid-disagree\endcsname{0.0}
\expandafter\def\csname val@rev.desc.d6.sspro.p1.C.tdiag.context-d1.share-agree-stable\endcsname{0.750}
\expandafter\def\csname cnt@rev.desc.d6.sspro.p1.C.tdiag.context-d1.share-agree-stable\endcsname{1}
\expandafter\def\csname pct@rev.desc.d6.sspro.p1.C.tdiag.context-d1.share-agree-stable\endcsname{75.0}
\expandafter\def\csname ppt@rev.desc.d6.sspro.p1.C.tdiag.context-d1.share-agree-stable\endcsname{$+$75.0}
\expandafter\def\csname val@rev.desc.d6.sspro.p1.C.tdiag.dual-d2.N-agree-stable\endcsname{36}
\expandafter\def\csname cnt@rev.desc.d6.sspro.p1.C.tdiag.dual-d2.N-agree-stable\endcsname{36}
\expandafter\def\csname pct@rev.desc.d6.sspro.p1.C.tdiag.dual-d2.N-agree-stable\endcsname{3600.0}
\expandafter\def\csname ppt@rev.desc.d6.sspro.p1.C.tdiag.dual-d2.N-agree-stable\endcsname{$+$3600.0}
\expandafter\def\csname val@rev.desc.d6.sspro.p1.C.tdiag.dual-d2.N-disagree\endcsname{9}
\expandafter\def\csname cnt@rev.desc.d6.sspro.p1.C.tdiag.dual-d2.N-disagree\endcsname{9}
\expandafter\def\csname pct@rev.desc.d6.sspro.p1.C.tdiag.dual-d2.N-disagree\endcsname{900.0}
\expandafter\def\csname ppt@rev.desc.d6.sspro.p1.C.tdiag.dual-d2.N-disagree\endcsname{$+$900.0}
\expandafter\def\csname val@rev.desc.d6.sspro.p1.C.tdiag.dual-d2.boot-resid.ci95.0\endcsname{0.023}
\expandafter\def\csname cnt@rev.desc.d6.sspro.p1.C.tdiag.dual-d2.boot-resid.ci95.0\endcsname{0}
\expandafter\def\csname pct@rev.desc.d6.sspro.p1.C.tdiag.dual-d2.boot-resid.ci95.0\endcsname{2.3}
\expandafter\def\csname ppt@rev.desc.d6.sspro.p1.C.tdiag.dual-d2.boot-resid.ci95.0\endcsname{$+$2.3}
\expandafter\def\csname val@rev.desc.d6.sspro.p1.C.tdiag.dual-d2.boot-resid.ci95.1\endcsname{0.241}
\expandafter\def\csname cnt@rev.desc.d6.sspro.p1.C.tdiag.dual-d2.boot-resid.ci95.1\endcsname{0}
\expandafter\def\csname pct@rev.desc.d6.sspro.p1.C.tdiag.dual-d2.boot-resid.ci95.1\endcsname{24.1}
\expandafter\def\csname ppt@rev.desc.d6.sspro.p1.C.tdiag.dual-d2.boot-resid.ci95.1\endcsname{$+$24.1}
\expandafter\def\csname val@rev.desc.d6.sspro.p1.C.tdiag.dual-d2.boot-resid.n-apps\endcsname{6}
\expandafter\def\csname cnt@rev.desc.d6.sspro.p1.C.tdiag.dual-d2.boot-resid.n-apps\endcsname{6}
\expandafter\def\csname pct@rev.desc.d6.sspro.p1.C.tdiag.dual-d2.boot-resid.n-apps\endcsname{600.0}
\expandafter\def\csname ppt@rev.desc.d6.sspro.p1.C.tdiag.dual-d2.boot-resid.n-apps\endcsname{$+$600.0}
\expandafter\def\csname val@rev.desc.d6.sspro.p1.C.tdiag.dual-d2.boot-resid.n-boot-valid\endcsname{4000}
\expandafter\def\csname cnt@rev.desc.d6.sspro.p1.C.tdiag.dual-d2.boot-resid.n-boot-valid\endcsname{4{,}000}
\expandafter\def\csname pct@rev.desc.d6.sspro.p1.C.tdiag.dual-d2.boot-resid.n-boot-valid\endcsname{400000.0}
\expandafter\def\csname ppt@rev.desc.d6.sspro.p1.C.tdiag.dual-d2.boot-resid.n-boot-valid\endcsname{$+$400000.0}
\expandafter\def\csname val@rev.desc.d6.sspro.p1.C.tdiag.dual-d2.boot-resid.value\endcsname{0.111}
\expandafter\def\csname cnt@rev.desc.d6.sspro.p1.C.tdiag.dual-d2.boot-resid.value\endcsname{0}
\expandafter\def\csname pct@rev.desc.d6.sspro.p1.C.tdiag.dual-d2.boot-resid.value\endcsname{11.1}
\expandafter\def\csname ppt@rev.desc.d6.sspro.p1.C.tdiag.dual-d2.boot-resid.value\endcsname{$+$11.1}
\expandafter\def\csname val@rev.desc.d6.sspro.p1.C.tdiag.dual-d2.boot-share.ci95.0\endcsname{0.521}
\expandafter\def\csname cnt@rev.desc.d6.sspro.p1.C.tdiag.dual-d2.boot-share.ci95.0\endcsname{1}
\expandafter\def\csname pct@rev.desc.d6.sspro.p1.C.tdiag.dual-d2.boot-share.ci95.0\endcsname{52.1}
\expandafter\def\csname ppt@rev.desc.d6.sspro.p1.C.tdiag.dual-d2.boot-share.ci95.0\endcsname{$+$52.1}
\expandafter\def\csname val@rev.desc.d6.sspro.p1.C.tdiag.dual-d2.boot-share.ci95.1\endcsname{0.938}
\expandafter\def\csname cnt@rev.desc.d6.sspro.p1.C.tdiag.dual-d2.boot-share.ci95.1\endcsname{1}
\expandafter\def\csname pct@rev.desc.d6.sspro.p1.C.tdiag.dual-d2.boot-share.ci95.1\endcsname{93.8}
\expandafter\def\csname ppt@rev.desc.d6.sspro.p1.C.tdiag.dual-d2.boot-share.ci95.1\endcsname{$+$93.8}
\expandafter\def\csname val@rev.desc.d6.sspro.p1.C.tdiag.dual-d2.boot-share.n-apps\endcsname{6}
\expandafter\def\csname cnt@rev.desc.d6.sspro.p1.C.tdiag.dual-d2.boot-share.n-apps\endcsname{6}
\expandafter\def\csname pct@rev.desc.d6.sspro.p1.C.tdiag.dual-d2.boot-share.n-apps\endcsname{600.0}
\expandafter\def\csname ppt@rev.desc.d6.sspro.p1.C.tdiag.dual-d2.boot-share.n-apps\endcsname{$+$600.0}
\expandafter\def\csname val@rev.desc.d6.sspro.p1.C.tdiag.dual-d2.boot-share.n-boot-valid\endcsname{4000}
\expandafter\def\csname cnt@rev.desc.d6.sspro.p1.C.tdiag.dual-d2.boot-share.n-boot-valid\endcsname{4{,}000}
\expandafter\def\csname pct@rev.desc.d6.sspro.p1.C.tdiag.dual-d2.boot-share.n-boot-valid\endcsname{400000.0}
\expandafter\def\csname ppt@rev.desc.d6.sspro.p1.C.tdiag.dual-d2.boot-share.n-boot-valid\endcsname{$+$400000.0}
\expandafter\def\csname val@rev.desc.d6.sspro.p1.C.tdiag.dual-d2.boot-share.value\endcsname{0.750}
\expandafter\def\csname cnt@rev.desc.d6.sspro.p1.C.tdiag.dual-d2.boot-share.value\endcsname{1}
\expandafter\def\csname pct@rev.desc.d6.sspro.p1.C.tdiag.dual-d2.boot-share.value\endcsname{75.0}
\expandafter\def\csname ppt@rev.desc.d6.sspro.p1.C.tdiag.dual-d2.boot-share.value\endcsname{$+$75.0}
\expandafter\def\csname val@rev.desc.d6.sspro.p1.C.tdiag.dual-d2.n\endcsname{48}
\expandafter\def\csname cnt@rev.desc.d6.sspro.p1.C.tdiag.dual-d2.n\endcsname{48}
\expandafter\def\csname pct@rev.desc.d6.sspro.p1.C.tdiag.dual-d2.n\endcsname{4800.0}
\expandafter\def\csname ppt@rev.desc.d6.sspro.p1.C.tdiag.dual-d2.n\endcsname{$+$4800.0}
\expandafter\def\csname val@rev.desc.d6.sspro.p1.C.tdiag.dual-d2.resid-agree-stable\endcsname{0.111}
\expandafter\def\csname cnt@rev.desc.d6.sspro.p1.C.tdiag.dual-d2.resid-agree-stable\endcsname{0}
\expandafter\def\csname pct@rev.desc.d6.sspro.p1.C.tdiag.dual-d2.resid-agree-stable\endcsname{11.1}
\expandafter\def\csname ppt@rev.desc.d6.sspro.p1.C.tdiag.dual-d2.resid-agree-stable\endcsname{$+$11.1}
\expandafter\def\csname val@rev.desc.d6.sspro.p1.C.tdiag.dual-d2.resid-disagree\endcsname{-0.111}
\expandafter\def\csname cnt@rev.desc.d6.sspro.p1.C.tdiag.dual-d2.resid-disagree\endcsname{0}
\expandafter\def\csname pct@rev.desc.d6.sspro.p1.C.tdiag.dual-d2.resid-disagree\endcsname{$-$11.1}
\expandafter\def\csname ppt@rev.desc.d6.sspro.p1.C.tdiag.dual-d2.resid-disagree\endcsname{$-$11.1}
\expandafter\def\csname val@rev.desc.d6.sspro.p1.C.tdiag.dual-d2.share-agree-stable\endcsname{0.750}
\expandafter\def\csname cnt@rev.desc.d6.sspro.p1.C.tdiag.dual-d2.share-agree-stable\endcsname{1}
\expandafter\def\csname pct@rev.desc.d6.sspro.p1.C.tdiag.dual-d2.share-agree-stable\endcsname{75.0}
\expandafter\def\csname ppt@rev.desc.d6.sspro.p1.C.tdiag.dual-d2.share-agree-stable\endcsname{$+$75.0}
\expandafter\def\csname val@rev.desc.d6.sspro.p1.C.tdiag.grid2-d3.N-agree-stable\endcsname{34}
\expandafter\def\csname cnt@rev.desc.d6.sspro.p1.C.tdiag.grid2-d3.N-agree-stable\endcsname{34}
\expandafter\def\csname pct@rev.desc.d6.sspro.p1.C.tdiag.grid2-d3.N-agree-stable\endcsname{3400.0}
\expandafter\def\csname ppt@rev.desc.d6.sspro.p1.C.tdiag.grid2-d3.N-agree-stable\endcsname{$+$3400.0}
\expandafter\def\csname val@rev.desc.d6.sspro.p1.C.tdiag.grid2-d3.N-disagree\endcsname{12}
\expandafter\def\csname cnt@rev.desc.d6.sspro.p1.C.tdiag.grid2-d3.N-disagree\endcsname{12}
\expandafter\def\csname pct@rev.desc.d6.sspro.p1.C.tdiag.grid2-d3.N-disagree\endcsname{1200.0}
\expandafter\def\csname ppt@rev.desc.d6.sspro.p1.C.tdiag.grid2-d3.N-disagree\endcsname{$+$1200.0}
\expandafter\def\csname val@rev.desc.d6.sspro.p1.C.tdiag.grid2-d3.boot-resid.ci95.0\endcsname{-0.129}
\expandafter\def\csname cnt@rev.desc.d6.sspro.p1.C.tdiag.grid2-d3.boot-resid.ci95.0\endcsname{0}
\expandafter\def\csname pct@rev.desc.d6.sspro.p1.C.tdiag.grid2-d3.boot-resid.ci95.0\endcsname{$-$12.9}
\expandafter\def\csname ppt@rev.desc.d6.sspro.p1.C.tdiag.grid2-d3.boot-resid.ci95.0\endcsname{$-$12.9}
\expandafter\def\csname val@rev.desc.d6.sspro.p1.C.tdiag.grid2-d3.boot-resid.ci95.1\endcsname{0.000}
\expandafter\def\csname cnt@rev.desc.d6.sspro.p1.C.tdiag.grid2-d3.boot-resid.ci95.1\endcsname{0}
\expandafter\def\csname pct@rev.desc.d6.sspro.p1.C.tdiag.grid2-d3.boot-resid.ci95.1\endcsname{0.0}
\expandafter\def\csname ppt@rev.desc.d6.sspro.p1.C.tdiag.grid2-d3.boot-resid.ci95.1\endcsname{0.0}
\expandafter\def\csname val@rev.desc.d6.sspro.p1.C.tdiag.grid2-d3.boot-resid.n-apps\endcsname{6}
\expandafter\def\csname cnt@rev.desc.d6.sspro.p1.C.tdiag.grid2-d3.boot-resid.n-apps\endcsname{6}
\expandafter\def\csname pct@rev.desc.d6.sspro.p1.C.tdiag.grid2-d3.boot-resid.n-apps\endcsname{600.0}
\expandafter\def\csname ppt@rev.desc.d6.sspro.p1.C.tdiag.grid2-d3.boot-resid.n-apps\endcsname{$+$600.0}
\expandafter\def\csname val@rev.desc.d6.sspro.p1.C.tdiag.grid2-d3.boot-resid.n-boot-valid\endcsname{4000}
\expandafter\def\csname cnt@rev.desc.d6.sspro.p1.C.tdiag.grid2-d3.boot-resid.n-boot-valid\endcsname{4{,}000}
\expandafter\def\csname pct@rev.desc.d6.sspro.p1.C.tdiag.grid2-d3.boot-resid.n-boot-valid\endcsname{400000.0}
\expandafter\def\csname ppt@rev.desc.d6.sspro.p1.C.tdiag.grid2-d3.boot-resid.n-boot-valid\endcsname{$+$400000.0}
\expandafter\def\csname val@rev.desc.d6.sspro.p1.C.tdiag.grid2-d3.boot-resid.value\endcsname{-0.059}
\expandafter\def\csname cnt@rev.desc.d6.sspro.p1.C.tdiag.grid2-d3.boot-resid.value\endcsname{0}
\expandafter\def\csname pct@rev.desc.d6.sspro.p1.C.tdiag.grid2-d3.boot-resid.value\endcsname{$-$5.9}
\expandafter\def\csname ppt@rev.desc.d6.sspro.p1.C.tdiag.grid2-d3.boot-resid.value\endcsname{$-$5.9}
\expandafter\def\csname val@rev.desc.d6.sspro.p1.C.tdiag.grid2-d3.boot-share.ci95.0\endcsname{0.542}
\expandafter\def\csname cnt@rev.desc.d6.sspro.p1.C.tdiag.grid2-d3.boot-share.ci95.0\endcsname{1}
\expandafter\def\csname pct@rev.desc.d6.sspro.p1.C.tdiag.grid2-d3.boot-share.ci95.0\endcsname{54.2}
\expandafter\def\csname ppt@rev.desc.d6.sspro.p1.C.tdiag.grid2-d3.boot-share.ci95.0\endcsname{$+$54.2}
\expandafter\def\csname val@rev.desc.d6.sspro.p1.C.tdiag.grid2-d3.boot-share.ci95.1\endcsname{0.854}
\expandafter\def\csname cnt@rev.desc.d6.sspro.p1.C.tdiag.grid2-d3.boot-share.ci95.1\endcsname{1}
\expandafter\def\csname pct@rev.desc.d6.sspro.p1.C.tdiag.grid2-d3.boot-share.ci95.1\endcsname{85.4}
\expandafter\def\csname ppt@rev.desc.d6.sspro.p1.C.tdiag.grid2-d3.boot-share.ci95.1\endcsname{$+$85.4}
\expandafter\def\csname val@rev.desc.d6.sspro.p1.C.tdiag.grid2-d3.boot-share.n-apps\endcsname{6}
\expandafter\def\csname cnt@rev.desc.d6.sspro.p1.C.tdiag.grid2-d3.boot-share.n-apps\endcsname{6}
\expandafter\def\csname pct@rev.desc.d6.sspro.p1.C.tdiag.grid2-d3.boot-share.n-apps\endcsname{600.0}
\expandafter\def\csname ppt@rev.desc.d6.sspro.p1.C.tdiag.grid2-d3.boot-share.n-apps\endcsname{$+$600.0}
\expandafter\def\csname val@rev.desc.d6.sspro.p1.C.tdiag.grid2-d3.boot-share.n-boot-valid\endcsname{4000}
\expandafter\def\csname cnt@rev.desc.d6.sspro.p1.C.tdiag.grid2-d3.boot-share.n-boot-valid\endcsname{4{,}000}
\expandafter\def\csname pct@rev.desc.d6.sspro.p1.C.tdiag.grid2-d3.boot-share.n-boot-valid\endcsname{400000.0}
\expandafter\def\csname ppt@rev.desc.d6.sspro.p1.C.tdiag.grid2-d3.boot-share.n-boot-valid\endcsname{$+$400000.0}
\expandafter\def\csname val@rev.desc.d6.sspro.p1.C.tdiag.grid2-d3.boot-share.value\endcsname{0.708}
\expandafter\def\csname cnt@rev.desc.d6.sspro.p1.C.tdiag.grid2-d3.boot-share.value\endcsname{1}
\expandafter\def\csname pct@rev.desc.d6.sspro.p1.C.tdiag.grid2-d3.boot-share.value\endcsname{70.8}
\expandafter\def\csname ppt@rev.desc.d6.sspro.p1.C.tdiag.grid2-d3.boot-share.value\endcsname{$+$70.8}
\expandafter\def\csname val@rev.desc.d6.sspro.p1.C.tdiag.grid2-d3.n\endcsname{48}
\expandafter\def\csname cnt@rev.desc.d6.sspro.p1.C.tdiag.grid2-d3.n\endcsname{48}
\expandafter\def\csname pct@rev.desc.d6.sspro.p1.C.tdiag.grid2-d3.n\endcsname{4800.0}
\expandafter\def\csname ppt@rev.desc.d6.sspro.p1.C.tdiag.grid2-d3.n\endcsname{$+$4800.0}
\expandafter\def\csname val@rev.desc.d6.sspro.p1.C.tdiag.grid2-d3.resid-agree-stable\endcsname{-0.059}
\expandafter\def\csname cnt@rev.desc.d6.sspro.p1.C.tdiag.grid2-d3.resid-agree-stable\endcsname{0}
\expandafter\def\csname pct@rev.desc.d6.sspro.p1.C.tdiag.grid2-d3.resid-agree-stable\endcsname{$-$5.9}
\expandafter\def\csname ppt@rev.desc.d6.sspro.p1.C.tdiag.grid2-d3.resid-agree-stable\endcsname{$-$5.9}
\expandafter\def\csname val@rev.desc.d6.sspro.p1.C.tdiag.grid2-d3.resid-disagree\endcsname{0.167}
\expandafter\def\csname cnt@rev.desc.d6.sspro.p1.C.tdiag.grid2-d3.resid-disagree\endcsname{0}
\expandafter\def\csname pct@rev.desc.d6.sspro.p1.C.tdiag.grid2-d3.resid-disagree\endcsname{16.7}
\expandafter\def\csname ppt@rev.desc.d6.sspro.p1.C.tdiag.grid2-d3.resid-disagree\endcsname{$+$16.7}
\expandafter\def\csname val@rev.desc.d6.sspro.p1.C.tdiag.grid2-d3.share-agree-stable\endcsname{0.708}
\expandafter\def\csname cnt@rev.desc.d6.sspro.p1.C.tdiag.grid2-d3.share-agree-stable\endcsname{1}
\expandafter\def\csname pct@rev.desc.d6.sspro.p1.C.tdiag.grid2-d3.share-agree-stable\endcsname{70.8}
\expandafter\def\csname ppt@rev.desc.d6.sspro.p1.C.tdiag.grid2-d3.share-agree-stable\endcsname{$+$70.8}
\expandafter\def\csname val@rev.desc.d6.sspro.p1.C.tdiag.zoomclick-d3.N-agree-stable\endcsname{37}
\expandafter\def\csname cnt@rev.desc.d6.sspro.p1.C.tdiag.zoomclick-d3.N-agree-stable\endcsname{37}
\expandafter\def\csname pct@rev.desc.d6.sspro.p1.C.tdiag.zoomclick-d3.N-agree-stable\endcsname{3700.0}
\expandafter\def\csname ppt@rev.desc.d6.sspro.p1.C.tdiag.zoomclick-d3.N-agree-stable\endcsname{$+$3700.0}
\expandafter\def\csname val@rev.desc.d6.sspro.p1.C.tdiag.zoomclick-d3.N-disagree\endcsname{9}
\expandafter\def\csname cnt@rev.desc.d6.sspro.p1.C.tdiag.zoomclick-d3.N-disagree\endcsname{9}
\expandafter\def\csname pct@rev.desc.d6.sspro.p1.C.tdiag.zoomclick-d3.N-disagree\endcsname{900.0}
\expandafter\def\csname ppt@rev.desc.d6.sspro.p1.C.tdiag.zoomclick-d3.N-disagree\endcsname{$+$900.0}
\expandafter\def\csname val@rev.desc.d6.sspro.p1.C.tdiag.zoomclick-d3.boot-resid.ci95.0\endcsname{0.000}
\expandafter\def\csname cnt@rev.desc.d6.sspro.p1.C.tdiag.zoomclick-d3.boot-resid.ci95.0\endcsname{0}
\expandafter\def\csname pct@rev.desc.d6.sspro.p1.C.tdiag.zoomclick-d3.boot-resid.ci95.0\endcsname{0.0}
\expandafter\def\csname ppt@rev.desc.d6.sspro.p1.C.tdiag.zoomclick-d3.boot-resid.ci95.0\endcsname{0.0}
\expandafter\def\csname val@rev.desc.d6.sspro.p1.C.tdiag.zoomclick-d3.boot-resid.ci95.1\endcsname{0.000}
\expandafter\def\csname cnt@rev.desc.d6.sspro.p1.C.tdiag.zoomclick-d3.boot-resid.ci95.1\endcsname{0}
\expandafter\def\csname pct@rev.desc.d6.sspro.p1.C.tdiag.zoomclick-d3.boot-resid.ci95.1\endcsname{0.0}
\expandafter\def\csname ppt@rev.desc.d6.sspro.p1.C.tdiag.zoomclick-d3.boot-resid.ci95.1\endcsname{0.0}
\expandafter\def\csname val@rev.desc.d6.sspro.p1.C.tdiag.zoomclick-d3.boot-resid.n-apps\endcsname{6}
\expandafter\def\csname cnt@rev.desc.d6.sspro.p1.C.tdiag.zoomclick-d3.boot-resid.n-apps\endcsname{6}
\expandafter\def\csname pct@rev.desc.d6.sspro.p1.C.tdiag.zoomclick-d3.boot-resid.n-apps\endcsname{600.0}
\expandafter\def\csname ppt@rev.desc.d6.sspro.p1.C.tdiag.zoomclick-d3.boot-resid.n-apps\endcsname{$+$600.0}
\expandafter\def\csname val@rev.desc.d6.sspro.p1.C.tdiag.zoomclick-d3.boot-resid.n-boot-valid\endcsname{4000}
\expandafter\def\csname cnt@rev.desc.d6.sspro.p1.C.tdiag.zoomclick-d3.boot-resid.n-boot-valid\endcsname{4{,}000}
\expandafter\def\csname pct@rev.desc.d6.sspro.p1.C.tdiag.zoomclick-d3.boot-resid.n-boot-valid\endcsname{400000.0}
\expandafter\def\csname ppt@rev.desc.d6.sspro.p1.C.tdiag.zoomclick-d3.boot-resid.n-boot-valid\endcsname{$+$400000.0}
\expandafter\def\csname val@rev.desc.d6.sspro.p1.C.tdiag.zoomclick-d3.boot-resid.value\endcsname{0.000}
\expandafter\def\csname cnt@rev.desc.d6.sspro.p1.C.tdiag.zoomclick-d3.boot-resid.value\endcsname{0}
\expandafter\def\csname pct@rev.desc.d6.sspro.p1.C.tdiag.zoomclick-d3.boot-resid.value\endcsname{0.0}
\expandafter\def\csname ppt@rev.desc.d6.sspro.p1.C.tdiag.zoomclick-d3.boot-resid.value\endcsname{0.0}
\expandafter\def\csname val@rev.desc.d6.sspro.p1.C.tdiag.zoomclick-d3.boot-share.ci95.0\endcsname{0.583}
\expandafter\def\csname cnt@rev.desc.d6.sspro.p1.C.tdiag.zoomclick-d3.boot-share.ci95.0\endcsname{1}
\expandafter\def\csname pct@rev.desc.d6.sspro.p1.C.tdiag.zoomclick-d3.boot-share.ci95.0\endcsname{58.3}
\expandafter\def\csname ppt@rev.desc.d6.sspro.p1.C.tdiag.zoomclick-d3.boot-share.ci95.0\endcsname{$+$58.3}
\expandafter\def\csname val@rev.desc.d6.sspro.p1.C.tdiag.zoomclick-d3.boot-share.ci95.1\endcsname{0.938}
\expandafter\def\csname cnt@rev.desc.d6.sspro.p1.C.tdiag.zoomclick-d3.boot-share.ci95.1\endcsname{1}
\expandafter\def\csname pct@rev.desc.d6.sspro.p1.C.tdiag.zoomclick-d3.boot-share.ci95.1\endcsname{93.8}
\expandafter\def\csname ppt@rev.desc.d6.sspro.p1.C.tdiag.zoomclick-d3.boot-share.ci95.1\endcsname{$+$93.8}
\expandafter\def\csname val@rev.desc.d6.sspro.p1.C.tdiag.zoomclick-d3.boot-share.n-apps\endcsname{6}
\expandafter\def\csname cnt@rev.desc.d6.sspro.p1.C.tdiag.zoomclick-d3.boot-share.n-apps\endcsname{6}
\expandafter\def\csname pct@rev.desc.d6.sspro.p1.C.tdiag.zoomclick-d3.boot-share.n-apps\endcsname{600.0}
\expandafter\def\csname ppt@rev.desc.d6.sspro.p1.C.tdiag.zoomclick-d3.boot-share.n-apps\endcsname{$+$600.0}
\expandafter\def\csname val@rev.desc.d6.sspro.p1.C.tdiag.zoomclick-d3.boot-share.n-boot-valid\endcsname{4000}
\expandafter\def\csname cnt@rev.desc.d6.sspro.p1.C.tdiag.zoomclick-d3.boot-share.n-boot-valid\endcsname{4{,}000}
\expandafter\def\csname pct@rev.desc.d6.sspro.p1.C.tdiag.zoomclick-d3.boot-share.n-boot-valid\endcsname{400000.0}
\expandafter\def\csname ppt@rev.desc.d6.sspro.p1.C.tdiag.zoomclick-d3.boot-share.n-boot-valid\endcsname{$+$400000.0}
\expandafter\def\csname val@rev.desc.d6.sspro.p1.C.tdiag.zoomclick-d3.boot-share.value\endcsname{0.771}
\expandafter\def\csname cnt@rev.desc.d6.sspro.p1.C.tdiag.zoomclick-d3.boot-share.value\endcsname{1}
\expandafter\def\csname pct@rev.desc.d6.sspro.p1.C.tdiag.zoomclick-d3.boot-share.value\endcsname{77.1}
\expandafter\def\csname ppt@rev.desc.d6.sspro.p1.C.tdiag.zoomclick-d3.boot-share.value\endcsname{$+$77.1}
\expandafter\def\csname val@rev.desc.d6.sspro.p1.C.tdiag.zoomclick-d3.n\endcsname{48}
\expandafter\def\csname cnt@rev.desc.d6.sspro.p1.C.tdiag.zoomclick-d3.n\endcsname{48}
\expandafter\def\csname pct@rev.desc.d6.sspro.p1.C.tdiag.zoomclick-d3.n\endcsname{4800.0}
\expandafter\def\csname ppt@rev.desc.d6.sspro.p1.C.tdiag.zoomclick-d3.n\endcsname{$+$4800.0}
\expandafter\def\csname val@rev.desc.d6.sspro.p1.C.tdiag.zoomclick-d3.resid-agree-stable\endcsname{0.000}
\expandafter\def\csname cnt@rev.desc.d6.sspro.p1.C.tdiag.zoomclick-d3.resid-agree-stable\endcsname{0}
\expandafter\def\csname pct@rev.desc.d6.sspro.p1.C.tdiag.zoomclick-d3.resid-agree-stable\endcsname{0.0}
\expandafter\def\csname ppt@rev.desc.d6.sspro.p1.C.tdiag.zoomclick-d3.resid-agree-stable\endcsname{0.0}
\expandafter\def\csname val@rev.desc.d6.sspro.p1.C.tdiag.zoomclick-d3.resid-disagree\endcsname{0.000}
\expandafter\def\csname cnt@rev.desc.d6.sspro.p1.C.tdiag.zoomclick-d3.resid-disagree\endcsname{0}
\expandafter\def\csname pct@rev.desc.d6.sspro.p1.C.tdiag.zoomclick-d3.resid-disagree\endcsname{0.0}
\expandafter\def\csname ppt@rev.desc.d6.sspro.p1.C.tdiag.zoomclick-d3.resid-disagree\endcsname{0.0}
\expandafter\def\csname val@rev.desc.d6.sspro.p1.C.tdiag.zoomclick-d3.share-agree-stable\endcsname{0.771}
\expandafter\def\csname cnt@rev.desc.d6.sspro.p1.C.tdiag.zoomclick-d3.share-agree-stable\endcsname{1}
\expandafter\def\csname pct@rev.desc.d6.sspro.p1.C.tdiag.zoomclick-d3.share-agree-stable\endcsname{77.1}
\expandafter\def\csname ppt@rev.desc.d6.sspro.p1.C.tdiag.zoomclick-d3.share-agree-stable\endcsname{$+$77.1}
\expandafter\def\csname val@rev.desc.d6.sspro.p2.C.box.catalog-pooled.N-agree-stable\endcsname{132}
\expandafter\def\csname cnt@rev.desc.d6.sspro.p2.C.box.catalog-pooled.N-agree-stable\endcsname{132}
\expandafter\def\csname pct@rev.desc.d6.sspro.p2.C.box.catalog-pooled.N-agree-stable\endcsname{13200.0}
\expandafter\def\csname ppt@rev.desc.d6.sspro.p2.C.box.catalog-pooled.N-agree-stable\endcsname{$+$13200.0}
\expandafter\def\csname val@rev.desc.d6.sspro.p2.C.box.catalog-pooled.N-disagree\endcsname{56}
\expandafter\def\csname cnt@rev.desc.d6.sspro.p2.C.box.catalog-pooled.N-disagree\endcsname{56}
\expandafter\def\csname pct@rev.desc.d6.sspro.p2.C.box.catalog-pooled.N-disagree\endcsname{5600.0}
\expandafter\def\csname ppt@rev.desc.d6.sspro.p2.C.box.catalog-pooled.N-disagree\endcsname{$+$5600.0}
\expandafter\def\csname val@rev.desc.d6.sspro.p2.C.box.catalog-pooled.boot-resid.ci95.0\endcsname{-0.075}
\expandafter\def\csname cnt@rev.desc.d6.sspro.p2.C.box.catalog-pooled.boot-resid.ci95.0\endcsname{0}
\expandafter\def\csname pct@rev.desc.d6.sspro.p2.C.box.catalog-pooled.boot-resid.ci95.0\endcsname{$-$7.5}
\expandafter\def\csname ppt@rev.desc.d6.sspro.p2.C.box.catalog-pooled.boot-resid.ci95.0\endcsname{$-$7.5}
\expandafter\def\csname val@rev.desc.d6.sspro.p2.C.box.catalog-pooled.boot-resid.ci95.1\endcsname{-0.007}
\expandafter\def\csname cnt@rev.desc.d6.sspro.p2.C.box.catalog-pooled.boot-resid.ci95.1\endcsname{0}
\expandafter\def\csname pct@rev.desc.d6.sspro.p2.C.box.catalog-pooled.boot-resid.ci95.1\endcsname{$-$0.7}
\expandafter\def\csname ppt@rev.desc.d6.sspro.p2.C.box.catalog-pooled.boot-resid.ci95.1\endcsname{$-$0.7}
\expandafter\def\csname val@rev.desc.d6.sspro.p2.C.box.catalog-pooled.boot-resid.n-apps\endcsname{6}
\expandafter\def\csname cnt@rev.desc.d6.sspro.p2.C.box.catalog-pooled.boot-resid.n-apps\endcsname{6}
\expandafter\def\csname pct@rev.desc.d6.sspro.p2.C.box.catalog-pooled.boot-resid.n-apps\endcsname{600.0}
\expandafter\def\csname ppt@rev.desc.d6.sspro.p2.C.box.catalog-pooled.boot-resid.n-apps\endcsname{$+$600.0}
\expandafter\def\csname val@rev.desc.d6.sspro.p2.C.box.catalog-pooled.boot-resid.n-boot-valid\endcsname{4000}
\expandafter\def\csname cnt@rev.desc.d6.sspro.p2.C.box.catalog-pooled.boot-resid.n-boot-valid\endcsname{4{,}000}
\expandafter\def\csname pct@rev.desc.d6.sspro.p2.C.box.catalog-pooled.boot-resid.n-boot-valid\endcsname{400000.0}
\expandafter\def\csname ppt@rev.desc.d6.sspro.p2.C.box.catalog-pooled.boot-resid.n-boot-valid\endcsname{$+$400000.0}
\expandafter\def\csname val@rev.desc.d6.sspro.p2.C.box.catalog-pooled.boot-resid.value\endcsname{-0.038}
\expandafter\def\csname cnt@rev.desc.d6.sspro.p2.C.box.catalog-pooled.boot-resid.value\endcsname{0}
\expandafter\def\csname pct@rev.desc.d6.sspro.p2.C.box.catalog-pooled.boot-resid.value\endcsname{$-$3.8}
\expandafter\def\csname ppt@rev.desc.d6.sspro.p2.C.box.catalog-pooled.boot-resid.value\endcsname{$-$3.8}
\expandafter\def\csname val@rev.desc.d6.sspro.p2.C.box.catalog-pooled.boot-share.ci95.0\endcsname{0.500}
\expandafter\def\csname cnt@rev.desc.d6.sspro.p2.C.box.catalog-pooled.boot-share.ci95.0\endcsname{0}
\expandafter\def\csname pct@rev.desc.d6.sspro.p2.C.box.catalog-pooled.boot-share.ci95.0\endcsname{50.0}
\expandafter\def\csname ppt@rev.desc.d6.sspro.p2.C.box.catalog-pooled.boot-share.ci95.0\endcsname{$+$50.0}
\expandafter\def\csname val@rev.desc.d6.sspro.p2.C.box.catalog-pooled.boot-share.ci95.1\endcsname{0.823}
\expandafter\def\csname cnt@rev.desc.d6.sspro.p2.C.box.catalog-pooled.boot-share.ci95.1\endcsname{1}
\expandafter\def\csname pct@rev.desc.d6.sspro.p2.C.box.catalog-pooled.boot-share.ci95.1\endcsname{82.3}
\expandafter\def\csname ppt@rev.desc.d6.sspro.p2.C.box.catalog-pooled.boot-share.ci95.1\endcsname{$+$82.3}
\expandafter\def\csname val@rev.desc.d6.sspro.p2.C.box.catalog-pooled.boot-share.n-apps\endcsname{6}
\expandafter\def\csname cnt@rev.desc.d6.sspro.p2.C.box.catalog-pooled.boot-share.n-apps\endcsname{6}
\expandafter\def\csname pct@rev.desc.d6.sspro.p2.C.box.catalog-pooled.boot-share.n-apps\endcsname{600.0}
\expandafter\def\csname ppt@rev.desc.d6.sspro.p2.C.box.catalog-pooled.boot-share.n-apps\endcsname{$+$600.0}
\expandafter\def\csname val@rev.desc.d6.sspro.p2.C.box.catalog-pooled.boot-share.n-boot-valid\endcsname{4000}
\expandafter\def\csname cnt@rev.desc.d6.sspro.p2.C.box.catalog-pooled.boot-share.n-boot-valid\endcsname{4{,}000}
\expandafter\def\csname pct@rev.desc.d6.sspro.p2.C.box.catalog-pooled.boot-share.n-boot-valid\endcsname{400000.0}
\expandafter\def\csname ppt@rev.desc.d6.sspro.p2.C.box.catalog-pooled.boot-share.n-boot-valid\endcsname{$+$400000.0}
\expandafter\def\csname val@rev.desc.d6.sspro.p2.C.box.catalog-pooled.boot-share.value\endcsname{0.688}
\expandafter\def\csname cnt@rev.desc.d6.sspro.p2.C.box.catalog-pooled.boot-share.value\endcsname{1}
\expandafter\def\csname pct@rev.desc.d6.sspro.p2.C.box.catalog-pooled.boot-share.value\endcsname{68.8}
\expandafter\def\csname ppt@rev.desc.d6.sspro.p2.C.box.catalog-pooled.boot-share.value\endcsname{$+$68.8}
\expandafter\def\csname val@rev.desc.d6.sspro.p2.C.box.catalog-pooled.gap\endcsname{0.145}
\expandafter\def\csname cnt@rev.desc.d6.sspro.p2.C.box.catalog-pooled.gap\endcsname{0}
\expandafter\def\csname pct@rev.desc.d6.sspro.p2.C.box.catalog-pooled.gap\endcsname{14.5}
\expandafter\def\csname ppt@rev.desc.d6.sspro.p2.C.box.catalog-pooled.gap\endcsname{$+$14.5}
\expandafter\def\csname val@rev.desc.d6.sspro.p2.C.box.catalog-pooled.n\endcsname{192}
\expandafter\def\csname cnt@rev.desc.d6.sspro.p2.C.box.catalog-pooled.n\endcsname{192}
\expandafter\def\csname pct@rev.desc.d6.sspro.p2.C.box.catalog-pooled.n\endcsname{19200.0}
\expandafter\def\csname ppt@rev.desc.d6.sspro.p2.C.box.catalog-pooled.n\endcsname{$+$19200.0}
\expandafter\def\csname val@rev.desc.d6.sspro.p2.C.box.catalog-pooled.resid-agree-stable\endcsname{-0.038}
\expandafter\def\csname cnt@rev.desc.d6.sspro.p2.C.box.catalog-pooled.resid-agree-stable\endcsname{0}
\expandafter\def\csname pct@rev.desc.d6.sspro.p2.C.box.catalog-pooled.resid-agree-stable\endcsname{$-$3.8}
\expandafter\def\csname ppt@rev.desc.d6.sspro.p2.C.box.catalog-pooled.resid-agree-stable\endcsname{$-$3.8}
\expandafter\def\csname val@rev.desc.d6.sspro.p2.C.box.catalog-pooled.resid-disagree\endcsname{0.107}
\expandafter\def\csname cnt@rev.desc.d6.sspro.p2.C.box.catalog-pooled.resid-disagree\endcsname{0}
\expandafter\def\csname pct@rev.desc.d6.sspro.p2.C.box.catalog-pooled.resid-disagree\endcsname{10.7}
\expandafter\def\csname ppt@rev.desc.d6.sspro.p2.C.box.catalog-pooled.resid-disagree\endcsname{$+$10.7}
\expandafter\def\csname val@rev.desc.d6.sspro.p2.C.box.catalog-pooled.share-agree-stable\endcsname{0.688}
\expandafter\def\csname cnt@rev.desc.d6.sspro.p2.C.box.catalog-pooled.share-agree-stable\endcsname{1}
\expandafter\def\csname pct@rev.desc.d6.sspro.p2.C.box.catalog-pooled.share-agree-stable\endcsname{68.8}
\expandafter\def\csname ppt@rev.desc.d6.sspro.p2.C.box.catalog-pooled.share-agree-stable\endcsname{$+$68.8}
\expandafter\def\csname val@rev.desc.d6.sspro.p2.C.box.center-d1.N-agree-stable\endcsname{34}
\expandafter\def\csname cnt@rev.desc.d6.sspro.p2.C.box.center-d1.N-agree-stable\endcsname{34}
\expandafter\def\csname pct@rev.desc.d6.sspro.p2.C.box.center-d1.N-agree-stable\endcsname{3400.0}
\expandafter\def\csname ppt@rev.desc.d6.sspro.p2.C.box.center-d1.N-agree-stable\endcsname{$+$3400.0}
\expandafter\def\csname val@rev.desc.d6.sspro.p2.C.box.center-d1.N-disagree\endcsname{14}
\expandafter\def\csname cnt@rev.desc.d6.sspro.p2.C.box.center-d1.N-disagree\endcsname{14}
\expandafter\def\csname pct@rev.desc.d6.sspro.p2.C.box.center-d1.N-disagree\endcsname{1400.0}
\expandafter\def\csname ppt@rev.desc.d6.sspro.p2.C.box.center-d1.N-disagree\endcsname{$+$1400.0}
\expandafter\def\csname val@rev.desc.d6.sspro.p2.C.box.center-d1.boot-resid.ci95.0\endcsname{0.000}
\expandafter\def\csname cnt@rev.desc.d6.sspro.p2.C.box.center-d1.boot-resid.ci95.0\endcsname{0}
\expandafter\def\csname pct@rev.desc.d6.sspro.p2.C.box.center-d1.boot-resid.ci95.0\endcsname{0.0}
\expandafter\def\csname ppt@rev.desc.d6.sspro.p2.C.box.center-d1.boot-resid.ci95.0\endcsname{0.0}
\expandafter\def\csname val@rev.desc.d6.sspro.p2.C.box.center-d1.boot-resid.ci95.1\endcsname{0.000}
\expandafter\def\csname cnt@rev.desc.d6.sspro.p2.C.box.center-d1.boot-resid.ci95.1\endcsname{0}
\expandafter\def\csname pct@rev.desc.d6.sspro.p2.C.box.center-d1.boot-resid.ci95.1\endcsname{0.0}
\expandafter\def\csname ppt@rev.desc.d6.sspro.p2.C.box.center-d1.boot-resid.ci95.1\endcsname{0.0}
\expandafter\def\csname val@rev.desc.d6.sspro.p2.C.box.center-d1.boot-resid.n-apps\endcsname{6}
\expandafter\def\csname cnt@rev.desc.d6.sspro.p2.C.box.center-d1.boot-resid.n-apps\endcsname{6}
\expandafter\def\csname pct@rev.desc.d6.sspro.p2.C.box.center-d1.boot-resid.n-apps\endcsname{600.0}
\expandafter\def\csname ppt@rev.desc.d6.sspro.p2.C.box.center-d1.boot-resid.n-apps\endcsname{$+$600.0}
\expandafter\def\csname val@rev.desc.d6.sspro.p2.C.box.center-d1.boot-resid.n-boot-valid\endcsname{4000}
\expandafter\def\csname cnt@rev.desc.d6.sspro.p2.C.box.center-d1.boot-resid.n-boot-valid\endcsname{4{,}000}
\expandafter\def\csname pct@rev.desc.d6.sspro.p2.C.box.center-d1.boot-resid.n-boot-valid\endcsname{400000.0}
\expandafter\def\csname ppt@rev.desc.d6.sspro.p2.C.box.center-d1.boot-resid.n-boot-valid\endcsname{$+$400000.0}
\expandafter\def\csname val@rev.desc.d6.sspro.p2.C.box.center-d1.boot-resid.value\endcsname{0.000}
\expandafter\def\csname cnt@rev.desc.d6.sspro.p2.C.box.center-d1.boot-resid.value\endcsname{0}
\expandafter\def\csname pct@rev.desc.d6.sspro.p2.C.box.center-d1.boot-resid.value\endcsname{0.0}
\expandafter\def\csname ppt@rev.desc.d6.sspro.p2.C.box.center-d1.boot-resid.value\endcsname{0.0}
\expandafter\def\csname val@rev.desc.d6.sspro.p2.C.box.center-d1.boot-share.ci95.0\endcsname{0.521}
\expandafter\def\csname cnt@rev.desc.d6.sspro.p2.C.box.center-d1.boot-share.ci95.0\endcsname{1}
\expandafter\def\csname pct@rev.desc.d6.sspro.p2.C.box.center-d1.boot-share.ci95.0\endcsname{52.1}
\expandafter\def\csname ppt@rev.desc.d6.sspro.p2.C.box.center-d1.boot-share.ci95.0\endcsname{$+$52.1}
\expandafter\def\csname val@rev.desc.d6.sspro.p2.C.box.center-d1.boot-share.ci95.1\endcsname{0.854}
\expandafter\def\csname cnt@rev.desc.d6.sspro.p2.C.box.center-d1.boot-share.ci95.1\endcsname{1}
\expandafter\def\csname pct@rev.desc.d6.sspro.p2.C.box.center-d1.boot-share.ci95.1\endcsname{85.4}
\expandafter\def\csname ppt@rev.desc.d6.sspro.p2.C.box.center-d1.boot-share.ci95.1\endcsname{$+$85.4}
\expandafter\def\csname val@rev.desc.d6.sspro.p2.C.box.center-d1.boot-share.n-apps\endcsname{6}
\expandafter\def\csname cnt@rev.desc.d6.sspro.p2.C.box.center-d1.boot-share.n-apps\endcsname{6}
\expandafter\def\csname pct@rev.desc.d6.sspro.p2.C.box.center-d1.boot-share.n-apps\endcsname{600.0}
\expandafter\def\csname ppt@rev.desc.d6.sspro.p2.C.box.center-d1.boot-share.n-apps\endcsname{$+$600.0}
\expandafter\def\csname val@rev.desc.d6.sspro.p2.C.box.center-d1.boot-share.n-boot-valid\endcsname{4000}
\expandafter\def\csname cnt@rev.desc.d6.sspro.p2.C.box.center-d1.boot-share.n-boot-valid\endcsname{4{,}000}
\expandafter\def\csname pct@rev.desc.d6.sspro.p2.C.box.center-d1.boot-share.n-boot-valid\endcsname{400000.0}
\expandafter\def\csname ppt@rev.desc.d6.sspro.p2.C.box.center-d1.boot-share.n-boot-valid\endcsname{$+$400000.0}
\expandafter\def\csname val@rev.desc.d6.sspro.p2.C.box.center-d1.boot-share.value\endcsname{0.708}
\expandafter\def\csname cnt@rev.desc.d6.sspro.p2.C.box.center-d1.boot-share.value\endcsname{1}
\expandafter\def\csname pct@rev.desc.d6.sspro.p2.C.box.center-d1.boot-share.value\endcsname{70.8}
\expandafter\def\csname ppt@rev.desc.d6.sspro.p2.C.box.center-d1.boot-share.value\endcsname{$+$70.8}
\expandafter\def\csname val@rev.desc.d6.sspro.p2.C.box.center-d1.n\endcsname{48}
\expandafter\def\csname cnt@rev.desc.d6.sspro.p2.C.box.center-d1.n\endcsname{48}
\expandafter\def\csname pct@rev.desc.d6.sspro.p2.C.box.center-d1.n\endcsname{4800.0}
\expandafter\def\csname ppt@rev.desc.d6.sspro.p2.C.box.center-d1.n\endcsname{$+$4800.0}
\expandafter\def\csname val@rev.desc.d6.sspro.p2.C.box.center-d1.resid-agree-stable\endcsname{0.000}
\expandafter\def\csname cnt@rev.desc.d6.sspro.p2.C.box.center-d1.resid-agree-stable\endcsname{0}
\expandafter\def\csname pct@rev.desc.d6.sspro.p2.C.box.center-d1.resid-agree-stable\endcsname{0.0}
\expandafter\def\csname ppt@rev.desc.d6.sspro.p2.C.box.center-d1.resid-agree-stable\endcsname{0.0}
\expandafter\def\csname val@rev.desc.d6.sspro.p2.C.box.center-d1.resid-disagree\endcsname{0.071}
\expandafter\def\csname cnt@rev.desc.d6.sspro.p2.C.box.center-d1.resid-disagree\endcsname{0}
\expandafter\def\csname pct@rev.desc.d6.sspro.p2.C.box.center-d1.resid-disagree\endcsname{7.1}
\expandafter\def\csname ppt@rev.desc.d6.sspro.p2.C.box.center-d1.resid-disagree\endcsname{$+$7.1}
\expandafter\def\csname val@rev.desc.d6.sspro.p2.C.box.center-d1.share-agree-stable\endcsname{0.708}
\expandafter\def\csname cnt@rev.desc.d6.sspro.p2.C.box.center-d1.share-agree-stable\endcsname{1}
\expandafter\def\csname pct@rev.desc.d6.sspro.p2.C.box.center-d1.share-agree-stable\endcsname{70.8}
\expandafter\def\csname ppt@rev.desc.d6.sspro.p2.C.box.center-d1.share-agree-stable\endcsname{$+$70.8}
\expandafter\def\csname val@rev.desc.d6.sspro.p2.C.box.context-d1.N-agree-stable\endcsname{34}
\expandafter\def\csname cnt@rev.desc.d6.sspro.p2.C.box.context-d1.N-agree-stable\endcsname{34}
\expandafter\def\csname pct@rev.desc.d6.sspro.p2.C.box.context-d1.N-agree-stable\endcsname{3400.0}
\expandafter\def\csname ppt@rev.desc.d6.sspro.p2.C.box.context-d1.N-agree-stable\endcsname{$+$3400.0}
\expandafter\def\csname val@rev.desc.d6.sspro.p2.C.box.context-d1.N-disagree\endcsname{13}
\expandafter\def\csname cnt@rev.desc.d6.sspro.p2.C.box.context-d1.N-disagree\endcsname{13}
\expandafter\def\csname pct@rev.desc.d6.sspro.p2.C.box.context-d1.N-disagree\endcsname{1300.0}
\expandafter\def\csname ppt@rev.desc.d6.sspro.p2.C.box.context-d1.N-disagree\endcsname{$+$1300.0}
\expandafter\def\csname val@rev.desc.d6.sspro.p2.C.box.context-d1.boot-resid.ci95.0\endcsname{-0.088}
\expandafter\def\csname cnt@rev.desc.d6.sspro.p2.C.box.context-d1.boot-resid.ci95.0\endcsname{0}
\expandafter\def\csname pct@rev.desc.d6.sspro.p2.C.box.context-d1.boot-resid.ci95.0\endcsname{$-$8.8}
\expandafter\def\csname ppt@rev.desc.d6.sspro.p2.C.box.context-d1.boot-resid.ci95.0\endcsname{$-$8.8}
\expandafter\def\csname val@rev.desc.d6.sspro.p2.C.box.context-d1.boot-resid.ci95.1\endcsname{0.000}
\expandafter\def\csname cnt@rev.desc.d6.sspro.p2.C.box.context-d1.boot-resid.ci95.1\endcsname{0}
\expandafter\def\csname pct@rev.desc.d6.sspro.p2.C.box.context-d1.boot-resid.ci95.1\endcsname{0.0}
\expandafter\def\csname ppt@rev.desc.d6.sspro.p2.C.box.context-d1.boot-resid.ci95.1\endcsname{0.0}
\expandafter\def\csname val@rev.desc.d6.sspro.p2.C.box.context-d1.boot-resid.n-apps\endcsname{6}
\expandafter\def\csname cnt@rev.desc.d6.sspro.p2.C.box.context-d1.boot-resid.n-apps\endcsname{6}
\expandafter\def\csname pct@rev.desc.d6.sspro.p2.C.box.context-d1.boot-resid.n-apps\endcsname{600.0}
\expandafter\def\csname ppt@rev.desc.d6.sspro.p2.C.box.context-d1.boot-resid.n-apps\endcsname{$+$600.0}
\expandafter\def\csname val@rev.desc.d6.sspro.p2.C.box.context-d1.boot-resid.n-boot-valid\endcsname{4000}
\expandafter\def\csname cnt@rev.desc.d6.sspro.p2.C.box.context-d1.boot-resid.n-boot-valid\endcsname{4{,}000}
\expandafter\def\csname pct@rev.desc.d6.sspro.p2.C.box.context-d1.boot-resid.n-boot-valid\endcsname{400000.0}
\expandafter\def\csname ppt@rev.desc.d6.sspro.p2.C.box.context-d1.boot-resid.n-boot-valid\endcsname{$+$400000.0}
\expandafter\def\csname val@rev.desc.d6.sspro.p2.C.box.context-d1.boot-resid.value\endcsname{-0.029}
\expandafter\def\csname cnt@rev.desc.d6.sspro.p2.C.box.context-d1.boot-resid.value\endcsname{0}
\expandafter\def\csname pct@rev.desc.d6.sspro.p2.C.box.context-d1.boot-resid.value\endcsname{$-$2.9}
\expandafter\def\csname ppt@rev.desc.d6.sspro.p2.C.box.context-d1.boot-resid.value\endcsname{$-$2.9}
\expandafter\def\csname val@rev.desc.d6.sspro.p2.C.box.context-d1.boot-share.ci95.0\endcsname{0.521}
\expandafter\def\csname cnt@rev.desc.d6.sspro.p2.C.box.context-d1.boot-share.ci95.0\endcsname{1}
\expandafter\def\csname pct@rev.desc.d6.sspro.p2.C.box.context-d1.boot-share.ci95.0\endcsname{52.1}
\expandafter\def\csname ppt@rev.desc.d6.sspro.p2.C.box.context-d1.boot-share.ci95.0\endcsname{$+$52.1}
\expandafter\def\csname val@rev.desc.d6.sspro.p2.C.box.context-d1.boot-share.ci95.1\endcsname{0.854}
\expandafter\def\csname cnt@rev.desc.d6.sspro.p2.C.box.context-d1.boot-share.ci95.1\endcsname{1}
\expandafter\def\csname pct@rev.desc.d6.sspro.p2.C.box.context-d1.boot-share.ci95.1\endcsname{85.4}
\expandafter\def\csname ppt@rev.desc.d6.sspro.p2.C.box.context-d1.boot-share.ci95.1\endcsname{$+$85.4}
\expandafter\def\csname val@rev.desc.d6.sspro.p2.C.box.context-d1.boot-share.n-apps\endcsname{6}
\expandafter\def\csname cnt@rev.desc.d6.sspro.p2.C.box.context-d1.boot-share.n-apps\endcsname{6}
\expandafter\def\csname pct@rev.desc.d6.sspro.p2.C.box.context-d1.boot-share.n-apps\endcsname{600.0}
\expandafter\def\csname ppt@rev.desc.d6.sspro.p2.C.box.context-d1.boot-share.n-apps\endcsname{$+$600.0}
\expandafter\def\csname val@rev.desc.d6.sspro.p2.C.box.context-d1.boot-share.n-boot-valid\endcsname{4000}
\expandafter\def\csname cnt@rev.desc.d6.sspro.p2.C.box.context-d1.boot-share.n-boot-valid\endcsname{4{,}000}
\expandafter\def\csname pct@rev.desc.d6.sspro.p2.C.box.context-d1.boot-share.n-boot-valid\endcsname{400000.0}
\expandafter\def\csname ppt@rev.desc.d6.sspro.p2.C.box.context-d1.boot-share.n-boot-valid\endcsname{$+$400000.0}
\expandafter\def\csname val@rev.desc.d6.sspro.p2.C.box.context-d1.boot-share.value\endcsname{0.708}
\expandafter\def\csname cnt@rev.desc.d6.sspro.p2.C.box.context-d1.boot-share.value\endcsname{1}
\expandafter\def\csname pct@rev.desc.d6.sspro.p2.C.box.context-d1.boot-share.value\endcsname{70.8}
\expandafter\def\csname ppt@rev.desc.d6.sspro.p2.C.box.context-d1.boot-share.value\endcsname{$+$70.8}
\expandafter\def\csname val@rev.desc.d6.sspro.p2.C.box.context-d1.n\endcsname{48}
\expandafter\def\csname cnt@rev.desc.d6.sspro.p2.C.box.context-d1.n\endcsname{48}
\expandafter\def\csname pct@rev.desc.d6.sspro.p2.C.box.context-d1.n\endcsname{4800.0}
\expandafter\def\csname ppt@rev.desc.d6.sspro.p2.C.box.context-d1.n\endcsname{$+$4800.0}
\expandafter\def\csname val@rev.desc.d6.sspro.p2.C.box.context-d1.resid-agree-stable\endcsname{-0.029}
\expandafter\def\csname cnt@rev.desc.d6.sspro.p2.C.box.context-d1.resid-agree-stable\endcsname{0}
\expandafter\def\csname pct@rev.desc.d6.sspro.p2.C.box.context-d1.resid-agree-stable\endcsname{$-$2.9}
\expandafter\def\csname ppt@rev.desc.d6.sspro.p2.C.box.context-d1.resid-agree-stable\endcsname{$-$2.9}
\expandafter\def\csname val@rev.desc.d6.sspro.p2.C.box.context-d1.resid-disagree\endcsname{0.077}
\expandafter\def\csname cnt@rev.desc.d6.sspro.p2.C.box.context-d1.resid-disagree\endcsname{0}
\expandafter\def\csname pct@rev.desc.d6.sspro.p2.C.box.context-d1.resid-disagree\endcsname{7.7}
\expandafter\def\csname ppt@rev.desc.d6.sspro.p2.C.box.context-d1.resid-disagree\endcsname{$+$7.7}
\expandafter\def\csname val@rev.desc.d6.sspro.p2.C.box.context-d1.share-agree-stable\endcsname{0.708}
\expandafter\def\csname cnt@rev.desc.d6.sspro.p2.C.box.context-d1.share-agree-stable\endcsname{1}
\expandafter\def\csname pct@rev.desc.d6.sspro.p2.C.box.context-d1.share-agree-stable\endcsname{70.8}
\expandafter\def\csname ppt@rev.desc.d6.sspro.p2.C.box.context-d1.share-agree-stable\endcsname{$+$70.8}
\expandafter\def\csname val@rev.desc.d6.sspro.p2.C.box.dual-d2.N-agree-stable\endcsname{31}
\expandafter\def\csname cnt@rev.desc.d6.sspro.p2.C.box.dual-d2.N-agree-stable\endcsname{31}
\expandafter\def\csname pct@rev.desc.d6.sspro.p2.C.box.dual-d2.N-agree-stable\endcsname{3100.0}
\expandafter\def\csname ppt@rev.desc.d6.sspro.p2.C.box.dual-d2.N-agree-stable\endcsname{$+$3100.0}
\expandafter\def\csname val@rev.desc.d6.sspro.p2.C.box.dual-d2.N-disagree\endcsname{17}
\expandafter\def\csname cnt@rev.desc.d6.sspro.p2.C.box.dual-d2.N-disagree\endcsname{17}
\expandafter\def\csname pct@rev.desc.d6.sspro.p2.C.box.dual-d2.N-disagree\endcsname{1700.0}
\expandafter\def\csname ppt@rev.desc.d6.sspro.p2.C.box.dual-d2.N-disagree\endcsname{$+$1700.0}
\expandafter\def\csname val@rev.desc.d6.sspro.p2.C.box.dual-d2.boot-resid.ci95.0\endcsname{-0.241}
\expandafter\def\csname cnt@rev.desc.d6.sspro.p2.C.box.dual-d2.boot-resid.ci95.0\endcsname{0}
\expandafter\def\csname pct@rev.desc.d6.sspro.p2.C.box.dual-d2.boot-resid.ci95.0\endcsname{$-$24.1}
\expandafter\def\csname ppt@rev.desc.d6.sspro.p2.C.box.dual-d2.boot-resid.ci95.0\endcsname{$-$24.1}
\expandafter\def\csname val@rev.desc.d6.sspro.p2.C.box.dual-d2.boot-resid.ci95.1\endcsname{0.000}
\expandafter\def\csname cnt@rev.desc.d6.sspro.p2.C.box.dual-d2.boot-resid.ci95.1\endcsname{0}
\expandafter\def\csname pct@rev.desc.d6.sspro.p2.C.box.dual-d2.boot-resid.ci95.1\endcsname{0.0}
\expandafter\def\csname ppt@rev.desc.d6.sspro.p2.C.box.dual-d2.boot-resid.ci95.1\endcsname{0.0}
\expandafter\def\csname val@rev.desc.d6.sspro.p2.C.box.dual-d2.boot-resid.n-apps\endcsname{6}
\expandafter\def\csname cnt@rev.desc.d6.sspro.p2.C.box.dual-d2.boot-resid.n-apps\endcsname{6}
\expandafter\def\csname pct@rev.desc.d6.sspro.p2.C.box.dual-d2.boot-resid.n-apps\endcsname{600.0}
\expandafter\def\csname ppt@rev.desc.d6.sspro.p2.C.box.dual-d2.boot-resid.n-apps\endcsname{$+$600.0}
\expandafter\def\csname val@rev.desc.d6.sspro.p2.C.box.dual-d2.boot-resid.n-boot-valid\endcsname{4000}
\expandafter\def\csname cnt@rev.desc.d6.sspro.p2.C.box.dual-d2.boot-resid.n-boot-valid\endcsname{4{,}000}
\expandafter\def\csname pct@rev.desc.d6.sspro.p2.C.box.dual-d2.boot-resid.n-boot-valid\endcsname{400000.0}
\expandafter\def\csname ppt@rev.desc.d6.sspro.p2.C.box.dual-d2.boot-resid.n-boot-valid\endcsname{$+$400000.0}
\expandafter\def\csname val@rev.desc.d6.sspro.p2.C.box.dual-d2.boot-resid.value\endcsname{-0.097}
\expandafter\def\csname cnt@rev.desc.d6.sspro.p2.C.box.dual-d2.boot-resid.value\endcsname{0}
\expandafter\def\csname pct@rev.desc.d6.sspro.p2.C.box.dual-d2.boot-resid.value\endcsname{$-$9.7}
\expandafter\def\csname ppt@rev.desc.d6.sspro.p2.C.box.dual-d2.boot-resid.value\endcsname{$-$9.7}
\expandafter\def\csname val@rev.desc.d6.sspro.p2.C.box.dual-d2.boot-share.ci95.0\endcsname{0.458}
\expandafter\def\csname cnt@rev.desc.d6.sspro.p2.C.box.dual-d2.boot-share.ci95.0\endcsname{0}
\expandafter\def\csname pct@rev.desc.d6.sspro.p2.C.box.dual-d2.boot-share.ci95.0\endcsname{45.8}
\expandafter\def\csname ppt@rev.desc.d6.sspro.p2.C.box.dual-d2.boot-share.ci95.0\endcsname{$+$45.8}
\expandafter\def\csname val@rev.desc.d6.sspro.p2.C.box.dual-d2.boot-share.ci95.1\endcsname{0.812}
\expandafter\def\csname cnt@rev.desc.d6.sspro.p2.C.box.dual-d2.boot-share.ci95.1\endcsname{1}
\expandafter\def\csname pct@rev.desc.d6.sspro.p2.C.box.dual-d2.boot-share.ci95.1\endcsname{81.2}
\expandafter\def\csname ppt@rev.desc.d6.sspro.p2.C.box.dual-d2.boot-share.ci95.1\endcsname{$+$81.2}
\expandafter\def\csname val@rev.desc.d6.sspro.p2.C.box.dual-d2.boot-share.n-apps\endcsname{6}
\expandafter\def\csname cnt@rev.desc.d6.sspro.p2.C.box.dual-d2.boot-share.n-apps\endcsname{6}
\expandafter\def\csname pct@rev.desc.d6.sspro.p2.C.box.dual-d2.boot-share.n-apps\endcsname{600.0}
\expandafter\def\csname ppt@rev.desc.d6.sspro.p2.C.box.dual-d2.boot-share.n-apps\endcsname{$+$600.0}
\expandafter\def\csname val@rev.desc.d6.sspro.p2.C.box.dual-d2.boot-share.n-boot-valid\endcsname{4000}
\expandafter\def\csname cnt@rev.desc.d6.sspro.p2.C.box.dual-d2.boot-share.n-boot-valid\endcsname{4{,}000}
\expandafter\def\csname pct@rev.desc.d6.sspro.p2.C.box.dual-d2.boot-share.n-boot-valid\endcsname{400000.0}
\expandafter\def\csname ppt@rev.desc.d6.sspro.p2.C.box.dual-d2.boot-share.n-boot-valid\endcsname{$+$400000.0}
\expandafter\def\csname val@rev.desc.d6.sspro.p2.C.box.dual-d2.boot-share.value\endcsname{0.646}
\expandafter\def\csname cnt@rev.desc.d6.sspro.p2.C.box.dual-d2.boot-share.value\endcsname{1}
\expandafter\def\csname pct@rev.desc.d6.sspro.p2.C.box.dual-d2.boot-share.value\endcsname{64.6}
\expandafter\def\csname ppt@rev.desc.d6.sspro.p2.C.box.dual-d2.boot-share.value\endcsname{$+$64.6}
\expandafter\def\csname val@rev.desc.d6.sspro.p2.C.box.dual-d2.n\endcsname{48}
\expandafter\def\csname cnt@rev.desc.d6.sspro.p2.C.box.dual-d2.n\endcsname{48}
\expandafter\def\csname pct@rev.desc.d6.sspro.p2.C.box.dual-d2.n\endcsname{4800.0}
\expandafter\def\csname ppt@rev.desc.d6.sspro.p2.C.box.dual-d2.n\endcsname{$+$4800.0}
\expandafter\def\csname val@rev.desc.d6.sspro.p2.C.box.dual-d2.resid-agree-stable\endcsname{-0.097}
\expandafter\def\csname cnt@rev.desc.d6.sspro.p2.C.box.dual-d2.resid-agree-stable\endcsname{0}
\expandafter\def\csname pct@rev.desc.d6.sspro.p2.C.box.dual-d2.resid-agree-stable\endcsname{$-$9.7}
\expandafter\def\csname ppt@rev.desc.d6.sspro.p2.C.box.dual-d2.resid-agree-stable\endcsname{$-$9.7}
\expandafter\def\csname val@rev.desc.d6.sspro.p2.C.box.dual-d2.resid-disagree\endcsname{0.235}
\expandafter\def\csname cnt@rev.desc.d6.sspro.p2.C.box.dual-d2.resid-disagree\endcsname{0}
\expandafter\def\csname pct@rev.desc.d6.sspro.p2.C.box.dual-d2.resid-disagree\endcsname{23.5}
\expandafter\def\csname ppt@rev.desc.d6.sspro.p2.C.box.dual-d2.resid-disagree\endcsname{$+$23.5}
\expandafter\def\csname val@rev.desc.d6.sspro.p2.C.box.dual-d2.share-agree-stable\endcsname{0.646}
\expandafter\def\csname cnt@rev.desc.d6.sspro.p2.C.box.dual-d2.share-agree-stable\endcsname{1}
\expandafter\def\csname pct@rev.desc.d6.sspro.p2.C.box.dual-d2.share-agree-stable\endcsname{64.6}
\expandafter\def\csname ppt@rev.desc.d6.sspro.p2.C.box.dual-d2.share-agree-stable\endcsname{$+$64.6}
\expandafter\def\csname val@rev.desc.d6.sspro.p2.C.box.grid2-d3.N-agree-stable\endcsname{33}
\expandafter\def\csname cnt@rev.desc.d6.sspro.p2.C.box.grid2-d3.N-agree-stable\endcsname{33}
\expandafter\def\csname pct@rev.desc.d6.sspro.p2.C.box.grid2-d3.N-agree-stable\endcsname{3300.0}
\expandafter\def\csname ppt@rev.desc.d6.sspro.p2.C.box.grid2-d3.N-agree-stable\endcsname{$+$3300.0}
\expandafter\def\csname val@rev.desc.d6.sspro.p2.C.box.grid2-d3.N-disagree\endcsname{12}
\expandafter\def\csname cnt@rev.desc.d6.sspro.p2.C.box.grid2-d3.N-disagree\endcsname{12}
\expandafter\def\csname pct@rev.desc.d6.sspro.p2.C.box.grid2-d3.N-disagree\endcsname{1200.0}
\expandafter\def\csname ppt@rev.desc.d6.sspro.p2.C.box.grid2-d3.N-disagree\endcsname{$+$1200.0}
\expandafter\def\csname val@rev.desc.d6.sspro.p2.C.box.grid2-d3.boot-resid.ci95.0\endcsname{-0.091}
\expandafter\def\csname cnt@rev.desc.d6.sspro.p2.C.box.grid2-d3.boot-resid.ci95.0\endcsname{0}
\expandafter\def\csname pct@rev.desc.d6.sspro.p2.C.box.grid2-d3.boot-resid.ci95.0\endcsname{$-$9.1}
\expandafter\def\csname ppt@rev.desc.d6.sspro.p2.C.box.grid2-d3.boot-resid.ci95.0\endcsname{$-$9.1}
\expandafter\def\csname val@rev.desc.d6.sspro.p2.C.box.grid2-d3.boot-resid.ci95.1\endcsname{0.000}
\expandafter\def\csname cnt@rev.desc.d6.sspro.p2.C.box.grid2-d3.boot-resid.ci95.1\endcsname{0}
\expandafter\def\csname pct@rev.desc.d6.sspro.p2.C.box.grid2-d3.boot-resid.ci95.1\endcsname{0.0}
\expandafter\def\csname ppt@rev.desc.d6.sspro.p2.C.box.grid2-d3.boot-resid.ci95.1\endcsname{0.0}
\expandafter\def\csname val@rev.desc.d6.sspro.p2.C.box.grid2-d3.boot-resid.n-apps\endcsname{6}
\expandafter\def\csname cnt@rev.desc.d6.sspro.p2.C.box.grid2-d3.boot-resid.n-apps\endcsname{6}
\expandafter\def\csname pct@rev.desc.d6.sspro.p2.C.box.grid2-d3.boot-resid.n-apps\endcsname{600.0}
\expandafter\def\csname ppt@rev.desc.d6.sspro.p2.C.box.grid2-d3.boot-resid.n-apps\endcsname{$+$600.0}
\expandafter\def\csname val@rev.desc.d6.sspro.p2.C.box.grid2-d3.boot-resid.n-boot-valid\endcsname{4000}
\expandafter\def\csname cnt@rev.desc.d6.sspro.p2.C.box.grid2-d3.boot-resid.n-boot-valid\endcsname{4{,}000}
\expandafter\def\csname pct@rev.desc.d6.sspro.p2.C.box.grid2-d3.boot-resid.n-boot-valid\endcsname{400000.0}
\expandafter\def\csname ppt@rev.desc.d6.sspro.p2.C.box.grid2-d3.boot-resid.n-boot-valid\endcsname{$+$400000.0}
\expandafter\def\csname val@rev.desc.d6.sspro.p2.C.box.grid2-d3.boot-resid.value\endcsname{-0.030}
\expandafter\def\csname cnt@rev.desc.d6.sspro.p2.C.box.grid2-d3.boot-resid.value\endcsname{0}
\expandafter\def\csname pct@rev.desc.d6.sspro.p2.C.box.grid2-d3.boot-resid.value\endcsname{$-$3.0}
\expandafter\def\csname ppt@rev.desc.d6.sspro.p2.C.box.grid2-d3.boot-resid.value\endcsname{$-$3.0}
\expandafter\def\csname val@rev.desc.d6.sspro.p2.C.box.grid2-d3.boot-share.ci95.0\endcsname{0.500}
\expandafter\def\csname cnt@rev.desc.d6.sspro.p2.C.box.grid2-d3.boot-share.ci95.0\endcsname{0}
\expandafter\def\csname pct@rev.desc.d6.sspro.p2.C.box.grid2-d3.boot-share.ci95.0\endcsname{50.0}
\expandafter\def\csname ppt@rev.desc.d6.sspro.p2.C.box.grid2-d3.boot-share.ci95.0\endcsname{$+$50.0}
\expandafter\def\csname val@rev.desc.d6.sspro.p2.C.box.grid2-d3.boot-share.ci95.1\endcsname{0.833}
\expandafter\def\csname cnt@rev.desc.d6.sspro.p2.C.box.grid2-d3.boot-share.ci95.1\endcsname{1}
\expandafter\def\csname pct@rev.desc.d6.sspro.p2.C.box.grid2-d3.boot-share.ci95.1\endcsname{83.3}
\expandafter\def\csname ppt@rev.desc.d6.sspro.p2.C.box.grid2-d3.boot-share.ci95.1\endcsname{$+$83.3}
\expandafter\def\csname val@rev.desc.d6.sspro.p2.C.box.grid2-d3.boot-share.n-apps\endcsname{6}
\expandafter\def\csname cnt@rev.desc.d6.sspro.p2.C.box.grid2-d3.boot-share.n-apps\endcsname{6}
\expandafter\def\csname pct@rev.desc.d6.sspro.p2.C.box.grid2-d3.boot-share.n-apps\endcsname{600.0}
\expandafter\def\csname ppt@rev.desc.d6.sspro.p2.C.box.grid2-d3.boot-share.n-apps\endcsname{$+$600.0}
\expandafter\def\csname val@rev.desc.d6.sspro.p2.C.box.grid2-d3.boot-share.n-boot-valid\endcsname{4000}
\expandafter\def\csname cnt@rev.desc.d6.sspro.p2.C.box.grid2-d3.boot-share.n-boot-valid\endcsname{4{,}000}
\expandafter\def\csname pct@rev.desc.d6.sspro.p2.C.box.grid2-d3.boot-share.n-boot-valid\endcsname{400000.0}
\expandafter\def\csname ppt@rev.desc.d6.sspro.p2.C.box.grid2-d3.boot-share.n-boot-valid\endcsname{$+$400000.0}
\expandafter\def\csname val@rev.desc.d6.sspro.p2.C.box.grid2-d3.boot-share.value\endcsname{0.688}
\expandafter\def\csname cnt@rev.desc.d6.sspro.p2.C.box.grid2-d3.boot-share.value\endcsname{1}
\expandafter\def\csname pct@rev.desc.d6.sspro.p2.C.box.grid2-d3.boot-share.value\endcsname{68.8}
\expandafter\def\csname ppt@rev.desc.d6.sspro.p2.C.box.grid2-d3.boot-share.value\endcsname{$+$68.8}
\expandafter\def\csname val@rev.desc.d6.sspro.p2.C.box.grid2-d3.n\endcsname{48}
\expandafter\def\csname cnt@rev.desc.d6.sspro.p2.C.box.grid2-d3.n\endcsname{48}
\expandafter\def\csname pct@rev.desc.d6.sspro.p2.C.box.grid2-d3.n\endcsname{4800.0}
\expandafter\def\csname ppt@rev.desc.d6.sspro.p2.C.box.grid2-d3.n\endcsname{$+$4800.0}
\expandafter\def\csname val@rev.desc.d6.sspro.p2.C.box.grid2-d3.resid-agree-stable\endcsname{-0.030}
\expandafter\def\csname cnt@rev.desc.d6.sspro.p2.C.box.grid2-d3.resid-agree-stable\endcsname{0}
\expandafter\def\csname pct@rev.desc.d6.sspro.p2.C.box.grid2-d3.resid-agree-stable\endcsname{$-$3.0}
\expandafter\def\csname ppt@rev.desc.d6.sspro.p2.C.box.grid2-d3.resid-agree-stable\endcsname{$-$3.0}
\expandafter\def\csname val@rev.desc.d6.sspro.p2.C.box.grid2-d3.resid-disagree\endcsname{0.000}
\expandafter\def\csname cnt@rev.desc.d6.sspro.p2.C.box.grid2-d3.resid-disagree\endcsname{0}
\expandafter\def\csname pct@rev.desc.d6.sspro.p2.C.box.grid2-d3.resid-disagree\endcsname{0.0}
\expandafter\def\csname ppt@rev.desc.d6.sspro.p2.C.box.grid2-d3.resid-disagree\endcsname{0.0}
\expandafter\def\csname val@rev.desc.d6.sspro.p2.C.box.grid2-d3.share-agree-stable\endcsname{0.688}
\expandafter\def\csname cnt@rev.desc.d6.sspro.p2.C.box.grid2-d3.share-agree-stable\endcsname{1}
\expandafter\def\csname pct@rev.desc.d6.sspro.p2.C.box.grid2-d3.share-agree-stable\endcsname{68.8}
\expandafter\def\csname ppt@rev.desc.d6.sspro.p2.C.box.grid2-d3.share-agree-stable\endcsname{$+$68.8}
\expandafter\def\csname val@rev.desc.d6.sspro.p2.C.box.zoomclick-d3.N-agree-stable\endcsname{34}
\expandafter\def\csname cnt@rev.desc.d6.sspro.p2.C.box.zoomclick-d3.N-agree-stable\endcsname{34}
\expandafter\def\csname pct@rev.desc.d6.sspro.p2.C.box.zoomclick-d3.N-agree-stable\endcsname{3400.0}
\expandafter\def\csname ppt@rev.desc.d6.sspro.p2.C.box.zoomclick-d3.N-agree-stable\endcsname{$+$3400.0}
\expandafter\def\csname val@rev.desc.d6.sspro.p2.C.box.zoomclick-d3.N-disagree\endcsname{13}
\expandafter\def\csname cnt@rev.desc.d6.sspro.p2.C.box.zoomclick-d3.N-disagree\endcsname{13}
\expandafter\def\csname pct@rev.desc.d6.sspro.p2.C.box.zoomclick-d3.N-disagree\endcsname{1300.0}
\expandafter\def\csname ppt@rev.desc.d6.sspro.p2.C.box.zoomclick-d3.N-disagree\endcsname{$+$1300.0}
\expandafter\def\csname val@rev.desc.d6.sspro.p2.C.box.zoomclick-d3.boot-resid.ci95.0\endcsname{0.000}
\expandafter\def\csname cnt@rev.desc.d6.sspro.p2.C.box.zoomclick-d3.boot-resid.ci95.0\endcsname{0}
\expandafter\def\csname pct@rev.desc.d6.sspro.p2.C.box.zoomclick-d3.boot-resid.ci95.0\endcsname{0.0}
\expandafter\def\csname ppt@rev.desc.d6.sspro.p2.C.box.zoomclick-d3.boot-resid.ci95.0\endcsname{0.0}
\expandafter\def\csname val@rev.desc.d6.sspro.p2.C.box.zoomclick-d3.boot-resid.ci95.1\endcsname{0.000}
\expandafter\def\csname cnt@rev.desc.d6.sspro.p2.C.box.zoomclick-d3.boot-resid.ci95.1\endcsname{0}
\expandafter\def\csname pct@rev.desc.d6.sspro.p2.C.box.zoomclick-d3.boot-resid.ci95.1\endcsname{0.0}
\expandafter\def\csname ppt@rev.desc.d6.sspro.p2.C.box.zoomclick-d3.boot-resid.ci95.1\endcsname{0.0}
\expandafter\def\csname val@rev.desc.d6.sspro.p2.C.box.zoomclick-d3.boot-resid.n-apps\endcsname{6}
\expandafter\def\csname cnt@rev.desc.d6.sspro.p2.C.box.zoomclick-d3.boot-resid.n-apps\endcsname{6}
\expandafter\def\csname pct@rev.desc.d6.sspro.p2.C.box.zoomclick-d3.boot-resid.n-apps\endcsname{600.0}
\expandafter\def\csname ppt@rev.desc.d6.sspro.p2.C.box.zoomclick-d3.boot-resid.n-apps\endcsname{$+$600.0}
\expandafter\def\csname val@rev.desc.d6.sspro.p2.C.box.zoomclick-d3.boot-resid.n-boot-valid\endcsname{4000}
\expandafter\def\csname cnt@rev.desc.d6.sspro.p2.C.box.zoomclick-d3.boot-resid.n-boot-valid\endcsname{4{,}000}
\expandafter\def\csname pct@rev.desc.d6.sspro.p2.C.box.zoomclick-d3.boot-resid.n-boot-valid\endcsname{400000.0}
\expandafter\def\csname ppt@rev.desc.d6.sspro.p2.C.box.zoomclick-d3.boot-resid.n-boot-valid\endcsname{$+$400000.0}
\expandafter\def\csname val@rev.desc.d6.sspro.p2.C.box.zoomclick-d3.boot-resid.value\endcsname{0.000}
\expandafter\def\csname cnt@rev.desc.d6.sspro.p2.C.box.zoomclick-d3.boot-resid.value\endcsname{0}
\expandafter\def\csname pct@rev.desc.d6.sspro.p2.C.box.zoomclick-d3.boot-resid.value\endcsname{0.0}
\expandafter\def\csname ppt@rev.desc.d6.sspro.p2.C.box.zoomclick-d3.boot-resid.value\endcsname{0.0}
\expandafter\def\csname val@rev.desc.d6.sspro.p2.C.box.zoomclick-d3.boot-share.ci95.0\endcsname{0.458}
\expandafter\def\csname cnt@rev.desc.d6.sspro.p2.C.box.zoomclick-d3.boot-share.ci95.0\endcsname{0}
\expandafter\def\csname pct@rev.desc.d6.sspro.p2.C.box.zoomclick-d3.boot-share.ci95.0\endcsname{45.8}
\expandafter\def\csname ppt@rev.desc.d6.sspro.p2.C.box.zoomclick-d3.boot-share.ci95.0\endcsname{$+$45.8}
\expandafter\def\csname val@rev.desc.d6.sspro.p2.C.box.zoomclick-d3.boot-share.ci95.1\endcsname{0.854}
\expandafter\def\csname cnt@rev.desc.d6.sspro.p2.C.box.zoomclick-d3.boot-share.ci95.1\endcsname{1}
\expandafter\def\csname pct@rev.desc.d6.sspro.p2.C.box.zoomclick-d3.boot-share.ci95.1\endcsname{85.4}
\expandafter\def\csname ppt@rev.desc.d6.sspro.p2.C.box.zoomclick-d3.boot-share.ci95.1\endcsname{$+$85.4}
\expandafter\def\csname val@rev.desc.d6.sspro.p2.C.box.zoomclick-d3.boot-share.n-apps\endcsname{6}
\expandafter\def\csname cnt@rev.desc.d6.sspro.p2.C.box.zoomclick-d3.boot-share.n-apps\endcsname{6}
\expandafter\def\csname pct@rev.desc.d6.sspro.p2.C.box.zoomclick-d3.boot-share.n-apps\endcsname{600.0}
\expandafter\def\csname ppt@rev.desc.d6.sspro.p2.C.box.zoomclick-d3.boot-share.n-apps\endcsname{$+$600.0}
\expandafter\def\csname val@rev.desc.d6.sspro.p2.C.box.zoomclick-d3.boot-share.n-boot-valid\endcsname{4000}
\expandafter\def\csname cnt@rev.desc.d6.sspro.p2.C.box.zoomclick-d3.boot-share.n-boot-valid\endcsname{4{,}000}
\expandafter\def\csname pct@rev.desc.d6.sspro.p2.C.box.zoomclick-d3.boot-share.n-boot-valid\endcsname{400000.0}
\expandafter\def\csname ppt@rev.desc.d6.sspro.p2.C.box.zoomclick-d3.boot-share.n-boot-valid\endcsname{$+$400000.0}
\expandafter\def\csname val@rev.desc.d6.sspro.p2.C.box.zoomclick-d3.boot-share.value\endcsname{0.708}
\expandafter\def\csname cnt@rev.desc.d6.sspro.p2.C.box.zoomclick-d3.boot-share.value\endcsname{1}
\expandafter\def\csname pct@rev.desc.d6.sspro.p2.C.box.zoomclick-d3.boot-share.value\endcsname{70.8}
\expandafter\def\csname ppt@rev.desc.d6.sspro.p2.C.box.zoomclick-d3.boot-share.value\endcsname{$+$70.8}
\expandafter\def\csname val@rev.desc.d6.sspro.p2.C.box.zoomclick-d3.n\endcsname{48}
\expandafter\def\csname cnt@rev.desc.d6.sspro.p2.C.box.zoomclick-d3.n\endcsname{48}
\expandafter\def\csname pct@rev.desc.d6.sspro.p2.C.box.zoomclick-d3.n\endcsname{4800.0}
\expandafter\def\csname ppt@rev.desc.d6.sspro.p2.C.box.zoomclick-d3.n\endcsname{$+$4800.0}
\expandafter\def\csname val@rev.desc.d6.sspro.p2.C.box.zoomclick-d3.resid-agree-stable\endcsname{0.000}
\expandafter\def\csname cnt@rev.desc.d6.sspro.p2.C.box.zoomclick-d3.resid-agree-stable\endcsname{0}
\expandafter\def\csname pct@rev.desc.d6.sspro.p2.C.box.zoomclick-d3.resid-agree-stable\endcsname{0.0}
\expandafter\def\csname ppt@rev.desc.d6.sspro.p2.C.box.zoomclick-d3.resid-agree-stable\endcsname{0.0}
\expandafter\def\csname val@rev.desc.d6.sspro.p2.C.box.zoomclick-d3.resid-disagree\endcsname{0.154}
\expandafter\def\csname cnt@rev.desc.d6.sspro.p2.C.box.zoomclick-d3.resid-disagree\endcsname{0}
\expandafter\def\csname pct@rev.desc.d6.sspro.p2.C.box.zoomclick-d3.resid-disagree\endcsname{15.4}
\expandafter\def\csname ppt@rev.desc.d6.sspro.p2.C.box.zoomclick-d3.resid-disagree\endcsname{$+$15.4}
\expandafter\def\csname val@rev.desc.d6.sspro.p2.C.box.zoomclick-d3.share-agree-stable\endcsname{0.708}
\expandafter\def\csname cnt@rev.desc.d6.sspro.p2.C.box.zoomclick-d3.share-agree-stable\endcsname{1}
\expandafter\def\csname pct@rev.desc.d6.sspro.p2.C.box.zoomclick-d3.share-agree-stable\endcsname{70.8}
\expandafter\def\csname ppt@rev.desc.d6.sspro.p2.C.box.zoomclick-d3.share-agree-stable\endcsname{$+$70.8}
\expandafter\def\csname val@rev.desc.d6.sspro.p2.C.tdiag.catalog-pooled.N-agree-stable\endcsname{150}
\expandafter\def\csname cnt@rev.desc.d6.sspro.p2.C.tdiag.catalog-pooled.N-agree-stable\endcsname{150}
\expandafter\def\csname pct@rev.desc.d6.sspro.p2.C.tdiag.catalog-pooled.N-agree-stable\endcsname{15000.0}
\expandafter\def\csname ppt@rev.desc.d6.sspro.p2.C.tdiag.catalog-pooled.N-agree-stable\endcsname{$+$15000.0}
\expandafter\def\csname val@rev.desc.d6.sspro.p2.C.tdiag.catalog-pooled.N-disagree\endcsname{38}
\expandafter\def\csname cnt@rev.desc.d6.sspro.p2.C.tdiag.catalog-pooled.N-disagree\endcsname{38}
\expandafter\def\csname pct@rev.desc.d6.sspro.p2.C.tdiag.catalog-pooled.N-disagree\endcsname{3800.0}
\expandafter\def\csname ppt@rev.desc.d6.sspro.p2.C.tdiag.catalog-pooled.N-disagree\endcsname{$+$3800.0}
\expandafter\def\csname val@rev.desc.d6.sspro.p2.C.tdiag.catalog-pooled.boot-resid.ci95.0\endcsname{-0.049}
\expandafter\def\csname cnt@rev.desc.d6.sspro.p2.C.tdiag.catalog-pooled.boot-resid.ci95.0\endcsname{0}
\expandafter\def\csname pct@rev.desc.d6.sspro.p2.C.tdiag.catalog-pooled.boot-resid.ci95.0\endcsname{$-$4.9}
\expandafter\def\csname ppt@rev.desc.d6.sspro.p2.C.tdiag.catalog-pooled.boot-resid.ci95.0\endcsname{$-$4.9}
\expandafter\def\csname val@rev.desc.d6.sspro.p2.C.tdiag.catalog-pooled.boot-resid.ci95.1\endcsname{0.000}
\expandafter\def\csname cnt@rev.desc.d6.sspro.p2.C.tdiag.catalog-pooled.boot-resid.ci95.1\endcsname{0}
\expandafter\def\csname pct@rev.desc.d6.sspro.p2.C.tdiag.catalog-pooled.boot-resid.ci95.1\endcsname{0.0}
\expandafter\def\csname ppt@rev.desc.d6.sspro.p2.C.tdiag.catalog-pooled.boot-resid.ci95.1\endcsname{0.0}
\expandafter\def\csname val@rev.desc.d6.sspro.p2.C.tdiag.catalog-pooled.boot-resid.n-apps\endcsname{6}
\expandafter\def\csname cnt@rev.desc.d6.sspro.p2.C.tdiag.catalog-pooled.boot-resid.n-apps\endcsname{6}
\expandafter\def\csname pct@rev.desc.d6.sspro.p2.C.tdiag.catalog-pooled.boot-resid.n-apps\endcsname{600.0}
\expandafter\def\csname ppt@rev.desc.d6.sspro.p2.C.tdiag.catalog-pooled.boot-resid.n-apps\endcsname{$+$600.0}
\expandafter\def\csname val@rev.desc.d6.sspro.p2.C.tdiag.catalog-pooled.boot-resid.n-boot-valid\endcsname{4000}
\expandafter\def\csname cnt@rev.desc.d6.sspro.p2.C.tdiag.catalog-pooled.boot-resid.n-boot-valid\endcsname{4{,}000}
\expandafter\def\csname pct@rev.desc.d6.sspro.p2.C.tdiag.catalog-pooled.boot-resid.n-boot-valid\endcsname{400000.0}
\expandafter\def\csname ppt@rev.desc.d6.sspro.p2.C.tdiag.catalog-pooled.boot-resid.n-boot-valid\endcsname{$+$400000.0}
\expandafter\def\csname val@rev.desc.d6.sspro.p2.C.tdiag.catalog-pooled.boot-resid.value\endcsname{-0.020}
\expandafter\def\csname cnt@rev.desc.d6.sspro.p2.C.tdiag.catalog-pooled.boot-resid.value\endcsname{0}
\expandafter\def\csname pct@rev.desc.d6.sspro.p2.C.tdiag.catalog-pooled.boot-resid.value\endcsname{$-$2.0}
\expandafter\def\csname ppt@rev.desc.d6.sspro.p2.C.tdiag.catalog-pooled.boot-resid.value\endcsname{$-$2.0}
\expandafter\def\csname val@rev.desc.d6.sspro.p2.C.tdiag.catalog-pooled.boot-share.ci95.0\endcsname{0.562}
\expandafter\def\csname cnt@rev.desc.d6.sspro.p2.C.tdiag.catalog-pooled.boot-share.ci95.0\endcsname{1}
\expandafter\def\csname pct@rev.desc.d6.sspro.p2.C.tdiag.catalog-pooled.boot-share.ci95.0\endcsname{56.2}
\expandafter\def\csname ppt@rev.desc.d6.sspro.p2.C.tdiag.catalog-pooled.boot-share.ci95.0\endcsname{$+$56.2}
\expandafter\def\csname val@rev.desc.d6.sspro.p2.C.tdiag.catalog-pooled.boot-share.ci95.1\endcsname{0.927}
\expandafter\def\csname cnt@rev.desc.d6.sspro.p2.C.tdiag.catalog-pooled.boot-share.ci95.1\endcsname{1}
\expandafter\def\csname pct@rev.desc.d6.sspro.p2.C.tdiag.catalog-pooled.boot-share.ci95.1\endcsname{92.7}
\expandafter\def\csname ppt@rev.desc.d6.sspro.p2.C.tdiag.catalog-pooled.boot-share.ci95.1\endcsname{$+$92.7}
\expandafter\def\csname val@rev.desc.d6.sspro.p2.C.tdiag.catalog-pooled.boot-share.n-apps\endcsname{6}
\expandafter\def\csname cnt@rev.desc.d6.sspro.p2.C.tdiag.catalog-pooled.boot-share.n-apps\endcsname{6}
\expandafter\def\csname pct@rev.desc.d6.sspro.p2.C.tdiag.catalog-pooled.boot-share.n-apps\endcsname{600.0}
\expandafter\def\csname ppt@rev.desc.d6.sspro.p2.C.tdiag.catalog-pooled.boot-share.n-apps\endcsname{$+$600.0}
\expandafter\def\csname val@rev.desc.d6.sspro.p2.C.tdiag.catalog-pooled.boot-share.n-boot-valid\endcsname{4000}
\expandafter\def\csname cnt@rev.desc.d6.sspro.p2.C.tdiag.catalog-pooled.boot-share.n-boot-valid\endcsname{4{,}000}
\expandafter\def\csname pct@rev.desc.d6.sspro.p2.C.tdiag.catalog-pooled.boot-share.n-boot-valid\endcsname{400000.0}
\expandafter\def\csname ppt@rev.desc.d6.sspro.p2.C.tdiag.catalog-pooled.boot-share.n-boot-valid\endcsname{$+$400000.0}
\expandafter\def\csname val@rev.desc.d6.sspro.p2.C.tdiag.catalog-pooled.boot-share.value\endcsname{0.781}
\expandafter\def\csname cnt@rev.desc.d6.sspro.p2.C.tdiag.catalog-pooled.boot-share.value\endcsname{1}
\expandafter\def\csname pct@rev.desc.d6.sspro.p2.C.tdiag.catalog-pooled.boot-share.value\endcsname{78.1}
\expandafter\def\csname ppt@rev.desc.d6.sspro.p2.C.tdiag.catalog-pooled.boot-share.value\endcsname{$+$78.1}
\expandafter\def\csname val@rev.desc.d6.sspro.p2.C.tdiag.catalog-pooled.gap\endcsname{0.125}
\expandafter\def\csname cnt@rev.desc.d6.sspro.p2.C.tdiag.catalog-pooled.gap\endcsname{0}
\expandafter\def\csname pct@rev.desc.d6.sspro.p2.C.tdiag.catalog-pooled.gap\endcsname{12.5}
\expandafter\def\csname ppt@rev.desc.d6.sspro.p2.C.tdiag.catalog-pooled.gap\endcsname{$+$12.5}
\expandafter\def\csname val@rev.desc.d6.sspro.p2.C.tdiag.catalog-pooled.n\endcsname{192}
\expandafter\def\csname cnt@rev.desc.d6.sspro.p2.C.tdiag.catalog-pooled.n\endcsname{192}
\expandafter\def\csname pct@rev.desc.d6.sspro.p2.C.tdiag.catalog-pooled.n\endcsname{19200.0}
\expandafter\def\csname ppt@rev.desc.d6.sspro.p2.C.tdiag.catalog-pooled.n\endcsname{$+$19200.0}
\expandafter\def\csname val@rev.desc.d6.sspro.p2.C.tdiag.catalog-pooled.resid-agree-stable\endcsname{-0.020}
\expandafter\def\csname cnt@rev.desc.d6.sspro.p2.C.tdiag.catalog-pooled.resid-agree-stable\endcsname{0}
\expandafter\def\csname pct@rev.desc.d6.sspro.p2.C.tdiag.catalog-pooled.resid-agree-stable\endcsname{$-$2.0}
\expandafter\def\csname ppt@rev.desc.d6.sspro.p2.C.tdiag.catalog-pooled.resid-agree-stable\endcsname{$-$2.0}
\expandafter\def\csname val@rev.desc.d6.sspro.p2.C.tdiag.catalog-pooled.resid-disagree\endcsname{0.105}
\expandafter\def\csname cnt@rev.desc.d6.sspro.p2.C.tdiag.catalog-pooled.resid-disagree\endcsname{0}
\expandafter\def\csname pct@rev.desc.d6.sspro.p2.C.tdiag.catalog-pooled.resid-disagree\endcsname{10.5}
\expandafter\def\csname ppt@rev.desc.d6.sspro.p2.C.tdiag.catalog-pooled.resid-disagree\endcsname{$+$10.5}
\expandafter\def\csname val@rev.desc.d6.sspro.p2.C.tdiag.catalog-pooled.share-agree-stable\endcsname{0.781}
\expandafter\def\csname cnt@rev.desc.d6.sspro.p2.C.tdiag.catalog-pooled.share-agree-stable\endcsname{1}
\expandafter\def\csname pct@rev.desc.d6.sspro.p2.C.tdiag.catalog-pooled.share-agree-stable\endcsname{78.1}
\expandafter\def\csname ppt@rev.desc.d6.sspro.p2.C.tdiag.catalog-pooled.share-agree-stable\endcsname{$+$78.1}
\expandafter\def\csname val@rev.desc.d6.sspro.p2.C.tdiag.center-d1.N-agree-stable\endcsname{39}
\expandafter\def\csname cnt@rev.desc.d6.sspro.p2.C.tdiag.center-d1.N-agree-stable\endcsname{39}
\expandafter\def\csname pct@rev.desc.d6.sspro.p2.C.tdiag.center-d1.N-agree-stable\endcsname{3900.0}
\expandafter\def\csname ppt@rev.desc.d6.sspro.p2.C.tdiag.center-d1.N-agree-stable\endcsname{$+$3900.0}
\expandafter\def\csname val@rev.desc.d6.sspro.p2.C.tdiag.center-d1.N-disagree\endcsname{9}
\expandafter\def\csname cnt@rev.desc.d6.sspro.p2.C.tdiag.center-d1.N-disagree\endcsname{9}
\expandafter\def\csname pct@rev.desc.d6.sspro.p2.C.tdiag.center-d1.N-disagree\endcsname{900.0}
\expandafter\def\csname ppt@rev.desc.d6.sspro.p2.C.tdiag.center-d1.N-disagree\endcsname{$+$900.0}
\expandafter\def\csname val@rev.desc.d6.sspro.p2.C.tdiag.center-d1.boot-resid.ci95.0\endcsname{0.000}
\expandafter\def\csname cnt@rev.desc.d6.sspro.p2.C.tdiag.center-d1.boot-resid.ci95.0\endcsname{0}
\expandafter\def\csname pct@rev.desc.d6.sspro.p2.C.tdiag.center-d1.boot-resid.ci95.0\endcsname{0.0}
\expandafter\def\csname ppt@rev.desc.d6.sspro.p2.C.tdiag.center-d1.boot-resid.ci95.0\endcsname{0.0}
\expandafter\def\csname val@rev.desc.d6.sspro.p2.C.tdiag.center-d1.boot-resid.ci95.1\endcsname{0.079}
\expandafter\def\csname cnt@rev.desc.d6.sspro.p2.C.tdiag.center-d1.boot-resid.ci95.1\endcsname{0}
\expandafter\def\csname pct@rev.desc.d6.sspro.p2.C.tdiag.center-d1.boot-resid.ci95.1\endcsname{7.9}
\expandafter\def\csname ppt@rev.desc.d6.sspro.p2.C.tdiag.center-d1.boot-resid.ci95.1\endcsname{$+$7.9}
\expandafter\def\csname val@rev.desc.d6.sspro.p2.C.tdiag.center-d1.boot-resid.n-apps\endcsname{6}
\expandafter\def\csname cnt@rev.desc.d6.sspro.p2.C.tdiag.center-d1.boot-resid.n-apps\endcsname{6}
\expandafter\def\csname pct@rev.desc.d6.sspro.p2.C.tdiag.center-d1.boot-resid.n-apps\endcsname{600.0}
\expandafter\def\csname ppt@rev.desc.d6.sspro.p2.C.tdiag.center-d1.boot-resid.n-apps\endcsname{$+$600.0}
\expandafter\def\csname val@rev.desc.d6.sspro.p2.C.tdiag.center-d1.boot-resid.n-boot-valid\endcsname{4000}
\expandafter\def\csname cnt@rev.desc.d6.sspro.p2.C.tdiag.center-d1.boot-resid.n-boot-valid\endcsname{4{,}000}
\expandafter\def\csname pct@rev.desc.d6.sspro.p2.C.tdiag.center-d1.boot-resid.n-boot-valid\endcsname{400000.0}
\expandafter\def\csname ppt@rev.desc.d6.sspro.p2.C.tdiag.center-d1.boot-resid.n-boot-valid\endcsname{$+$400000.0}
\expandafter\def\csname val@rev.desc.d6.sspro.p2.C.tdiag.center-d1.boot-resid.value\endcsname{0.026}
\expandafter\def\csname cnt@rev.desc.d6.sspro.p2.C.tdiag.center-d1.boot-resid.value\endcsname{0}
\expandafter\def\csname pct@rev.desc.d6.sspro.p2.C.tdiag.center-d1.boot-resid.value\endcsname{2.6}
\expandafter\def\csname ppt@rev.desc.d6.sspro.p2.C.tdiag.center-d1.boot-resid.value\endcsname{$+$2.6}
\expandafter\def\csname val@rev.desc.d6.sspro.p2.C.tdiag.center-d1.boot-share.ci95.0\endcsname{0.583}
\expandafter\def\csname cnt@rev.desc.d6.sspro.p2.C.tdiag.center-d1.boot-share.ci95.0\endcsname{1}
\expandafter\def\csname pct@rev.desc.d6.sspro.p2.C.tdiag.center-d1.boot-share.ci95.0\endcsname{58.3}
\expandafter\def\csname ppt@rev.desc.d6.sspro.p2.C.tdiag.center-d1.boot-share.ci95.0\endcsname{$+$58.3}
\expandafter\def\csname val@rev.desc.d6.sspro.p2.C.tdiag.center-d1.boot-share.ci95.1\endcsname{0.958}
\expandafter\def\csname cnt@rev.desc.d6.sspro.p2.C.tdiag.center-d1.boot-share.ci95.1\endcsname{1}
\expandafter\def\csname pct@rev.desc.d6.sspro.p2.C.tdiag.center-d1.boot-share.ci95.1\endcsname{95.8}
\expandafter\def\csname ppt@rev.desc.d6.sspro.p2.C.tdiag.center-d1.boot-share.ci95.1\endcsname{$+$95.8}
\expandafter\def\csname val@rev.desc.d6.sspro.p2.C.tdiag.center-d1.boot-share.n-apps\endcsname{6}
\expandafter\def\csname cnt@rev.desc.d6.sspro.p2.C.tdiag.center-d1.boot-share.n-apps\endcsname{6}
\expandafter\def\csname pct@rev.desc.d6.sspro.p2.C.tdiag.center-d1.boot-share.n-apps\endcsname{600.0}
\expandafter\def\csname ppt@rev.desc.d6.sspro.p2.C.tdiag.center-d1.boot-share.n-apps\endcsname{$+$600.0}
\expandafter\def\csname val@rev.desc.d6.sspro.p2.C.tdiag.center-d1.boot-share.n-boot-valid\endcsname{4000}
\expandafter\def\csname cnt@rev.desc.d6.sspro.p2.C.tdiag.center-d1.boot-share.n-boot-valid\endcsname{4{,}000}
\expandafter\def\csname pct@rev.desc.d6.sspro.p2.C.tdiag.center-d1.boot-share.n-boot-valid\endcsname{400000.0}
\expandafter\def\csname ppt@rev.desc.d6.sspro.p2.C.tdiag.center-d1.boot-share.n-boot-valid\endcsname{$+$400000.0}
\expandafter\def\csname val@rev.desc.d6.sspro.p2.C.tdiag.center-d1.boot-share.value\endcsname{0.812}
\expandafter\def\csname cnt@rev.desc.d6.sspro.p2.C.tdiag.center-d1.boot-share.value\endcsname{1}
\expandafter\def\csname pct@rev.desc.d6.sspro.p2.C.tdiag.center-d1.boot-share.value\endcsname{81.2}
\expandafter\def\csname ppt@rev.desc.d6.sspro.p2.C.tdiag.center-d1.boot-share.value\endcsname{$+$81.2}
\expandafter\def\csname val@rev.desc.d6.sspro.p2.C.tdiag.center-d1.n\endcsname{48}
\expandafter\def\csname cnt@rev.desc.d6.sspro.p2.C.tdiag.center-d1.n\endcsname{48}
\expandafter\def\csname pct@rev.desc.d6.sspro.p2.C.tdiag.center-d1.n\endcsname{4800.0}
\expandafter\def\csname ppt@rev.desc.d6.sspro.p2.C.tdiag.center-d1.n\endcsname{$+$4800.0}
\expandafter\def\csname val@rev.desc.d6.sspro.p2.C.tdiag.center-d1.resid-agree-stable\endcsname{0.026}
\expandafter\def\csname cnt@rev.desc.d6.sspro.p2.C.tdiag.center-d1.resid-agree-stable\endcsname{0}
\expandafter\def\csname pct@rev.desc.d6.sspro.p2.C.tdiag.center-d1.resid-agree-stable\endcsname{2.6}
\expandafter\def\csname ppt@rev.desc.d6.sspro.p2.C.tdiag.center-d1.resid-agree-stable\endcsname{$+$2.6}
\expandafter\def\csname val@rev.desc.d6.sspro.p2.C.tdiag.center-d1.resid-disagree\endcsname{0.000}
\expandafter\def\csname cnt@rev.desc.d6.sspro.p2.C.tdiag.center-d1.resid-disagree\endcsname{0}
\expandafter\def\csname pct@rev.desc.d6.sspro.p2.C.tdiag.center-d1.resid-disagree\endcsname{0.0}
\expandafter\def\csname ppt@rev.desc.d6.sspro.p2.C.tdiag.center-d1.resid-disagree\endcsname{0.0}
\expandafter\def\csname val@rev.desc.d6.sspro.p2.C.tdiag.center-d1.share-agree-stable\endcsname{0.812}
\expandafter\def\csname cnt@rev.desc.d6.sspro.p2.C.tdiag.center-d1.share-agree-stable\endcsname{1}
\expandafter\def\csname pct@rev.desc.d6.sspro.p2.C.tdiag.center-d1.share-agree-stable\endcsname{81.2}
\expandafter\def\csname ppt@rev.desc.d6.sspro.p2.C.tdiag.center-d1.share-agree-stable\endcsname{$+$81.2}
\expandafter\def\csname val@rev.desc.d6.sspro.p2.C.tdiag.context-d1.N-agree-stable\endcsname{38}
\expandafter\def\csname cnt@rev.desc.d6.sspro.p2.C.tdiag.context-d1.N-agree-stable\endcsname{38}
\expandafter\def\csname pct@rev.desc.d6.sspro.p2.C.tdiag.context-d1.N-agree-stable\endcsname{3800.0}
\expandafter\def\csname ppt@rev.desc.d6.sspro.p2.C.tdiag.context-d1.N-agree-stable\endcsname{$+$3800.0}
\expandafter\def\csname val@rev.desc.d6.sspro.p2.C.tdiag.context-d1.N-disagree\endcsname{9}
\expandafter\def\csname cnt@rev.desc.d6.sspro.p2.C.tdiag.context-d1.N-disagree\endcsname{9}
\expandafter\def\csname pct@rev.desc.d6.sspro.p2.C.tdiag.context-d1.N-disagree\endcsname{900.0}
\expandafter\def\csname ppt@rev.desc.d6.sspro.p2.C.tdiag.context-d1.N-disagree\endcsname{$+$900.0}
\expandafter\def\csname val@rev.desc.d6.sspro.p2.C.tdiag.context-d1.boot-resid.ci95.0\endcsname{0.000}
\expandafter\def\csname cnt@rev.desc.d6.sspro.p2.C.tdiag.context-d1.boot-resid.ci95.0\endcsname{0}
\expandafter\def\csname pct@rev.desc.d6.sspro.p2.C.tdiag.context-d1.boot-resid.ci95.0\endcsname{0.0}
\expandafter\def\csname ppt@rev.desc.d6.sspro.p2.C.tdiag.context-d1.boot-resid.ci95.0\endcsname{0.0}
\expandafter\def\csname val@rev.desc.d6.sspro.p2.C.tdiag.context-d1.boot-resid.ci95.1\endcsname{0.000}
\expandafter\def\csname cnt@rev.desc.d6.sspro.p2.C.tdiag.context-d1.boot-resid.ci95.1\endcsname{0}
\expandafter\def\csname pct@rev.desc.d6.sspro.p2.C.tdiag.context-d1.boot-resid.ci95.1\endcsname{0.0}
\expandafter\def\csname ppt@rev.desc.d6.sspro.p2.C.tdiag.context-d1.boot-resid.ci95.1\endcsname{0.0}
\expandafter\def\csname val@rev.desc.d6.sspro.p2.C.tdiag.context-d1.boot-resid.n-apps\endcsname{6}
\expandafter\def\csname cnt@rev.desc.d6.sspro.p2.C.tdiag.context-d1.boot-resid.n-apps\endcsname{6}
\expandafter\def\csname pct@rev.desc.d6.sspro.p2.C.tdiag.context-d1.boot-resid.n-apps\endcsname{600.0}
\expandafter\def\csname ppt@rev.desc.d6.sspro.p2.C.tdiag.context-d1.boot-resid.n-apps\endcsname{$+$600.0}
\expandafter\def\csname val@rev.desc.d6.sspro.p2.C.tdiag.context-d1.boot-resid.n-boot-valid\endcsname{4000}
\expandafter\def\csname cnt@rev.desc.d6.sspro.p2.C.tdiag.context-d1.boot-resid.n-boot-valid\endcsname{4{,}000}
\expandafter\def\csname pct@rev.desc.d6.sspro.p2.C.tdiag.context-d1.boot-resid.n-boot-valid\endcsname{400000.0}
\expandafter\def\csname ppt@rev.desc.d6.sspro.p2.C.tdiag.context-d1.boot-resid.n-boot-valid\endcsname{$+$400000.0}
\expandafter\def\csname val@rev.desc.d6.sspro.p2.C.tdiag.context-d1.boot-resid.value\endcsname{0.000}
\expandafter\def\csname cnt@rev.desc.d6.sspro.p2.C.tdiag.context-d1.boot-resid.value\endcsname{0}
\expandafter\def\csname pct@rev.desc.d6.sspro.p2.C.tdiag.context-d1.boot-resid.value\endcsname{0.0}
\expandafter\def\csname ppt@rev.desc.d6.sspro.p2.C.tdiag.context-d1.boot-resid.value\endcsname{0.0}
\expandafter\def\csname val@rev.desc.d6.sspro.p2.C.tdiag.context-d1.boot-share.ci95.0\endcsname{0.562}
\expandafter\def\csname cnt@rev.desc.d6.sspro.p2.C.tdiag.context-d1.boot-share.ci95.0\endcsname{1}
\expandafter\def\csname pct@rev.desc.d6.sspro.p2.C.tdiag.context-d1.boot-share.ci95.0\endcsname{56.2}
\expandafter\def\csname ppt@rev.desc.d6.sspro.p2.C.tdiag.context-d1.boot-share.ci95.0\endcsname{$+$56.2}
\expandafter\def\csname val@rev.desc.d6.sspro.p2.C.tdiag.context-d1.boot-share.ci95.1\endcsname{0.958}
\expandafter\def\csname cnt@rev.desc.d6.sspro.p2.C.tdiag.context-d1.boot-share.ci95.1\endcsname{1}
\expandafter\def\csname pct@rev.desc.d6.sspro.p2.C.tdiag.context-d1.boot-share.ci95.1\endcsname{95.8}
\expandafter\def\csname ppt@rev.desc.d6.sspro.p2.C.tdiag.context-d1.boot-share.ci95.1\endcsname{$+$95.8}
\expandafter\def\csname val@rev.desc.d6.sspro.p2.C.tdiag.context-d1.boot-share.n-apps\endcsname{6}
\expandafter\def\csname cnt@rev.desc.d6.sspro.p2.C.tdiag.context-d1.boot-share.n-apps\endcsname{6}
\expandafter\def\csname pct@rev.desc.d6.sspro.p2.C.tdiag.context-d1.boot-share.n-apps\endcsname{600.0}
\expandafter\def\csname ppt@rev.desc.d6.sspro.p2.C.tdiag.context-d1.boot-share.n-apps\endcsname{$+$600.0}
\expandafter\def\csname val@rev.desc.d6.sspro.p2.C.tdiag.context-d1.boot-share.n-boot-valid\endcsname{4000}
\expandafter\def\csname cnt@rev.desc.d6.sspro.p2.C.tdiag.context-d1.boot-share.n-boot-valid\endcsname{4{,}000}
\expandafter\def\csname pct@rev.desc.d6.sspro.p2.C.tdiag.context-d1.boot-share.n-boot-valid\endcsname{400000.0}
\expandafter\def\csname ppt@rev.desc.d6.sspro.p2.C.tdiag.context-d1.boot-share.n-boot-valid\endcsname{$+$400000.0}
\expandafter\def\csname val@rev.desc.d6.sspro.p2.C.tdiag.context-d1.boot-share.value\endcsname{0.792}
\expandafter\def\csname cnt@rev.desc.d6.sspro.p2.C.tdiag.context-d1.boot-share.value\endcsname{1}
\expandafter\def\csname pct@rev.desc.d6.sspro.p2.C.tdiag.context-d1.boot-share.value\endcsname{79.2}
\expandafter\def\csname ppt@rev.desc.d6.sspro.p2.C.tdiag.context-d1.boot-share.value\endcsname{$+$79.2}
\expandafter\def\csname val@rev.desc.d6.sspro.p2.C.tdiag.context-d1.n\endcsname{48}
\expandafter\def\csname cnt@rev.desc.d6.sspro.p2.C.tdiag.context-d1.n\endcsname{48}
\expandafter\def\csname pct@rev.desc.d6.sspro.p2.C.tdiag.context-d1.n\endcsname{4800.0}
\expandafter\def\csname ppt@rev.desc.d6.sspro.p2.C.tdiag.context-d1.n\endcsname{$+$4800.0}
\expandafter\def\csname val@rev.desc.d6.sspro.p2.C.tdiag.context-d1.resid-agree-stable\endcsname{0.000}
\expandafter\def\csname cnt@rev.desc.d6.sspro.p2.C.tdiag.context-d1.resid-agree-stable\endcsname{0}
\expandafter\def\csname pct@rev.desc.d6.sspro.p2.C.tdiag.context-d1.resid-agree-stable\endcsname{0.0}
\expandafter\def\csname ppt@rev.desc.d6.sspro.p2.C.tdiag.context-d1.resid-agree-stable\endcsname{0.0}
\expandafter\def\csname val@rev.desc.d6.sspro.p2.C.tdiag.context-d1.resid-disagree\endcsname{0.000}
\expandafter\def\csname cnt@rev.desc.d6.sspro.p2.C.tdiag.context-d1.resid-disagree\endcsname{0}
\expandafter\def\csname pct@rev.desc.d6.sspro.p2.C.tdiag.context-d1.resid-disagree\endcsname{0.0}
\expandafter\def\csname ppt@rev.desc.d6.sspro.p2.C.tdiag.context-d1.resid-disagree\endcsname{0.0}
\expandafter\def\csname val@rev.desc.d6.sspro.p2.C.tdiag.context-d1.share-agree-stable\endcsname{0.792}
\expandafter\def\csname cnt@rev.desc.d6.sspro.p2.C.tdiag.context-d1.share-agree-stable\endcsname{1}
\expandafter\def\csname pct@rev.desc.d6.sspro.p2.C.tdiag.context-d1.share-agree-stable\endcsname{79.2}
\expandafter\def\csname ppt@rev.desc.d6.sspro.p2.C.tdiag.context-d1.share-agree-stable\endcsname{$+$79.2}
\expandafter\def\csname val@rev.desc.d6.sspro.p2.C.tdiag.dual-d2.N-agree-stable\endcsname{36}
\expandafter\def\csname cnt@rev.desc.d6.sspro.p2.C.tdiag.dual-d2.N-agree-stable\endcsname{36}
\expandafter\def\csname pct@rev.desc.d6.sspro.p2.C.tdiag.dual-d2.N-agree-stable\endcsname{3600.0}
\expandafter\def\csname ppt@rev.desc.d6.sspro.p2.C.tdiag.dual-d2.N-agree-stable\endcsname{$+$3600.0}
\expandafter\def\csname val@rev.desc.d6.sspro.p2.C.tdiag.dual-d2.N-disagree\endcsname{12}
\expandafter\def\csname cnt@rev.desc.d6.sspro.p2.C.tdiag.dual-d2.N-disagree\endcsname{12}
\expandafter\def\csname pct@rev.desc.d6.sspro.p2.C.tdiag.dual-d2.N-disagree\endcsname{1200.0}
\expandafter\def\csname ppt@rev.desc.d6.sspro.p2.C.tdiag.dual-d2.N-disagree\endcsname{$+$1200.0}
\expandafter\def\csname val@rev.desc.d6.sspro.p2.C.tdiag.dual-d2.boot-resid.ci95.0\endcsname{-0.200}
\expandafter\def\csname cnt@rev.desc.d6.sspro.p2.C.tdiag.dual-d2.boot-resid.ci95.0\endcsname{0}
\expandafter\def\csname pct@rev.desc.d6.sspro.p2.C.tdiag.dual-d2.boot-resid.ci95.0\endcsname{$-$20.0}
\expandafter\def\csname ppt@rev.desc.d6.sspro.p2.C.tdiag.dual-d2.boot-resid.ci95.0\endcsname{$-$20.0}
\expandafter\def\csname val@rev.desc.d6.sspro.p2.C.tdiag.dual-d2.boot-resid.ci95.1\endcsname{0.000}
\expandafter\def\csname cnt@rev.desc.d6.sspro.p2.C.tdiag.dual-d2.boot-resid.ci95.1\endcsname{0}
\expandafter\def\csname pct@rev.desc.d6.sspro.p2.C.tdiag.dual-d2.boot-resid.ci95.1\endcsname{0.0}
\expandafter\def\csname ppt@rev.desc.d6.sspro.p2.C.tdiag.dual-d2.boot-resid.ci95.1\endcsname{0.0}
\expandafter\def\csname val@rev.desc.d6.sspro.p2.C.tdiag.dual-d2.boot-resid.n-apps\endcsname{6}
\expandafter\def\csname cnt@rev.desc.d6.sspro.p2.C.tdiag.dual-d2.boot-resid.n-apps\endcsname{6}
\expandafter\def\csname pct@rev.desc.d6.sspro.p2.C.tdiag.dual-d2.boot-resid.n-apps\endcsname{600.0}
\expandafter\def\csname ppt@rev.desc.d6.sspro.p2.C.tdiag.dual-d2.boot-resid.n-apps\endcsname{$+$600.0}
\expandafter\def\csname val@rev.desc.d6.sspro.p2.C.tdiag.dual-d2.boot-resid.n-boot-valid\endcsname{4000}
\expandafter\def\csname cnt@rev.desc.d6.sspro.p2.C.tdiag.dual-d2.boot-resid.n-boot-valid\endcsname{4{,}000}
\expandafter\def\csname pct@rev.desc.d6.sspro.p2.C.tdiag.dual-d2.boot-resid.n-boot-valid\endcsname{400000.0}
\expandafter\def\csname ppt@rev.desc.d6.sspro.p2.C.tdiag.dual-d2.boot-resid.n-boot-valid\endcsname{$+$400000.0}
\expandafter\def\csname val@rev.desc.d6.sspro.p2.C.tdiag.dual-d2.boot-resid.value\endcsname{-0.083}
\expandafter\def\csname cnt@rev.desc.d6.sspro.p2.C.tdiag.dual-d2.boot-resid.value\endcsname{0}
\expandafter\def\csname pct@rev.desc.d6.sspro.p2.C.tdiag.dual-d2.boot-resid.value\endcsname{$-$8.3}
\expandafter\def\csname ppt@rev.desc.d6.sspro.p2.C.tdiag.dual-d2.boot-resid.value\endcsname{$-$8.3}
\expandafter\def\csname val@rev.desc.d6.sspro.p2.C.tdiag.dual-d2.boot-share.ci95.0\endcsname{0.542}
\expandafter\def\csname cnt@rev.desc.d6.sspro.p2.C.tdiag.dual-d2.boot-share.ci95.0\endcsname{1}
\expandafter\def\csname pct@rev.desc.d6.sspro.p2.C.tdiag.dual-d2.boot-share.ci95.0\endcsname{54.2}
\expandafter\def\csname ppt@rev.desc.d6.sspro.p2.C.tdiag.dual-d2.boot-share.ci95.0\endcsname{$+$54.2}
\expandafter\def\csname val@rev.desc.d6.sspro.p2.C.tdiag.dual-d2.boot-share.ci95.1\endcsname{0.917}
\expandafter\def\csname cnt@rev.desc.d6.sspro.p2.C.tdiag.dual-d2.boot-share.ci95.1\endcsname{1}
\expandafter\def\csname pct@rev.desc.d6.sspro.p2.C.tdiag.dual-d2.boot-share.ci95.1\endcsname{91.7}
\expandafter\def\csname ppt@rev.desc.d6.sspro.p2.C.tdiag.dual-d2.boot-share.ci95.1\endcsname{$+$91.7}
\expandafter\def\csname val@rev.desc.d6.sspro.p2.C.tdiag.dual-d2.boot-share.n-apps\endcsname{6}
\expandafter\def\csname cnt@rev.desc.d6.sspro.p2.C.tdiag.dual-d2.boot-share.n-apps\endcsname{6}
\expandafter\def\csname pct@rev.desc.d6.sspro.p2.C.tdiag.dual-d2.boot-share.n-apps\endcsname{600.0}
\expandafter\def\csname ppt@rev.desc.d6.sspro.p2.C.tdiag.dual-d2.boot-share.n-apps\endcsname{$+$600.0}
\expandafter\def\csname val@rev.desc.d6.sspro.p2.C.tdiag.dual-d2.boot-share.n-boot-valid\endcsname{4000}
\expandafter\def\csname cnt@rev.desc.d6.sspro.p2.C.tdiag.dual-d2.boot-share.n-boot-valid\endcsname{4{,}000}
\expandafter\def\csname pct@rev.desc.d6.sspro.p2.C.tdiag.dual-d2.boot-share.n-boot-valid\endcsname{400000.0}
\expandafter\def\csname ppt@rev.desc.d6.sspro.p2.C.tdiag.dual-d2.boot-share.n-boot-valid\endcsname{$+$400000.0}
\expandafter\def\csname val@rev.desc.d6.sspro.p2.C.tdiag.dual-d2.boot-share.value\endcsname{0.750}
\expandafter\def\csname cnt@rev.desc.d6.sspro.p2.C.tdiag.dual-d2.boot-share.value\endcsname{1}
\expandafter\def\csname pct@rev.desc.d6.sspro.p2.C.tdiag.dual-d2.boot-share.value\endcsname{75.0}
\expandafter\def\csname ppt@rev.desc.d6.sspro.p2.C.tdiag.dual-d2.boot-share.value\endcsname{$+$75.0}
\expandafter\def\csname val@rev.desc.d6.sspro.p2.C.tdiag.dual-d2.n\endcsname{48}
\expandafter\def\csname cnt@rev.desc.d6.sspro.p2.C.tdiag.dual-d2.n\endcsname{48}
\expandafter\def\csname pct@rev.desc.d6.sspro.p2.C.tdiag.dual-d2.n\endcsname{4800.0}
\expandafter\def\csname ppt@rev.desc.d6.sspro.p2.C.tdiag.dual-d2.n\endcsname{$+$4800.0}
\expandafter\def\csname val@rev.desc.d6.sspro.p2.C.tdiag.dual-d2.resid-agree-stable\endcsname{-0.083}
\expandafter\def\csname cnt@rev.desc.d6.sspro.p2.C.tdiag.dual-d2.resid-agree-stable\endcsname{0}
\expandafter\def\csname pct@rev.desc.d6.sspro.p2.C.tdiag.dual-d2.resid-agree-stable\endcsname{$-$8.3}
\expandafter\def\csname ppt@rev.desc.d6.sspro.p2.C.tdiag.dual-d2.resid-agree-stable\endcsname{$-$8.3}
\expandafter\def\csname val@rev.desc.d6.sspro.p2.C.tdiag.dual-d2.resid-disagree\endcsname{0.333}
\expandafter\def\csname cnt@rev.desc.d6.sspro.p2.C.tdiag.dual-d2.resid-disagree\endcsname{0}
\expandafter\def\csname pct@rev.desc.d6.sspro.p2.C.tdiag.dual-d2.resid-disagree\endcsname{33.3}
\expandafter\def\csname ppt@rev.desc.d6.sspro.p2.C.tdiag.dual-d2.resid-disagree\endcsname{$+$33.3}
\expandafter\def\csname val@rev.desc.d6.sspro.p2.C.tdiag.dual-d2.share-agree-stable\endcsname{0.750}
\expandafter\def\csname cnt@rev.desc.d6.sspro.p2.C.tdiag.dual-d2.share-agree-stable\endcsname{1}
\expandafter\def\csname pct@rev.desc.d6.sspro.p2.C.tdiag.dual-d2.share-agree-stable\endcsname{75.0}
\expandafter\def\csname ppt@rev.desc.d6.sspro.p2.C.tdiag.dual-d2.share-agree-stable\endcsname{$+$75.0}
\expandafter\def\csname val@rev.desc.d6.sspro.p2.C.tdiag.grid2-d3.N-agree-stable\endcsname{37}
\expandafter\def\csname cnt@rev.desc.d6.sspro.p2.C.tdiag.grid2-d3.N-agree-stable\endcsname{37}
\expandafter\def\csname pct@rev.desc.d6.sspro.p2.C.tdiag.grid2-d3.N-agree-stable\endcsname{3700.0}
\expandafter\def\csname ppt@rev.desc.d6.sspro.p2.C.tdiag.grid2-d3.N-agree-stable\endcsname{$+$3700.0}
\expandafter\def\csname val@rev.desc.d6.sspro.p2.C.tdiag.grid2-d3.N-disagree\endcsname{8}
\expandafter\def\csname cnt@rev.desc.d6.sspro.p2.C.tdiag.grid2-d3.N-disagree\endcsname{8}
\expandafter\def\csname pct@rev.desc.d6.sspro.p2.C.tdiag.grid2-d3.N-disagree\endcsname{800.0}
\expandafter\def\csname ppt@rev.desc.d6.sspro.p2.C.tdiag.grid2-d3.N-disagree\endcsname{$+$800.0}
\expandafter\def\csname val@rev.desc.d6.sspro.p2.C.tdiag.grid2-d3.boot-resid.ci95.0\endcsname{-0.081}
\expandafter\def\csname cnt@rev.desc.d6.sspro.p2.C.tdiag.grid2-d3.boot-resid.ci95.0\endcsname{0}
\expandafter\def\csname pct@rev.desc.d6.sspro.p2.C.tdiag.grid2-d3.boot-resid.ci95.0\endcsname{$-$8.1}
\expandafter\def\csname ppt@rev.desc.d6.sspro.p2.C.tdiag.grid2-d3.boot-resid.ci95.0\endcsname{$-$8.1}
\expandafter\def\csname val@rev.desc.d6.sspro.p2.C.tdiag.grid2-d3.boot-resid.ci95.1\endcsname{0.000}
\expandafter\def\csname cnt@rev.desc.d6.sspro.p2.C.tdiag.grid2-d3.boot-resid.ci95.1\endcsname{0}
\expandafter\def\csname pct@rev.desc.d6.sspro.p2.C.tdiag.grid2-d3.boot-resid.ci95.1\endcsname{0.0}
\expandafter\def\csname ppt@rev.desc.d6.sspro.p2.C.tdiag.grid2-d3.boot-resid.ci95.1\endcsname{0.0}
\expandafter\def\csname val@rev.desc.d6.sspro.p2.C.tdiag.grid2-d3.boot-resid.n-apps\endcsname{6}
\expandafter\def\csname cnt@rev.desc.d6.sspro.p2.C.tdiag.grid2-d3.boot-resid.n-apps\endcsname{6}
\expandafter\def\csname pct@rev.desc.d6.sspro.p2.C.tdiag.grid2-d3.boot-resid.n-apps\endcsname{600.0}
\expandafter\def\csname ppt@rev.desc.d6.sspro.p2.C.tdiag.grid2-d3.boot-resid.n-apps\endcsname{$+$600.0}
\expandafter\def\csname val@rev.desc.d6.sspro.p2.C.tdiag.grid2-d3.boot-resid.n-boot-valid\endcsname{4000}
\expandafter\def\csname cnt@rev.desc.d6.sspro.p2.C.tdiag.grid2-d3.boot-resid.n-boot-valid\endcsname{4{,}000}
\expandafter\def\csname pct@rev.desc.d6.sspro.p2.C.tdiag.grid2-d3.boot-resid.n-boot-valid\endcsname{400000.0}
\expandafter\def\csname ppt@rev.desc.d6.sspro.p2.C.tdiag.grid2-d3.boot-resid.n-boot-valid\endcsname{$+$400000.0}
\expandafter\def\csname val@rev.desc.d6.sspro.p2.C.tdiag.grid2-d3.boot-resid.value\endcsname{-0.027}
\expandafter\def\csname cnt@rev.desc.d6.sspro.p2.C.tdiag.grid2-d3.boot-resid.value\endcsname{0}
\expandafter\def\csname pct@rev.desc.d6.sspro.p2.C.tdiag.grid2-d3.boot-resid.value\endcsname{$-$2.7}
\expandafter\def\csname ppt@rev.desc.d6.sspro.p2.C.tdiag.grid2-d3.boot-resid.value\endcsname{$-$2.7}
\expandafter\def\csname val@rev.desc.d6.sspro.p2.C.tdiag.grid2-d3.boot-share.ci95.0\endcsname{0.542}
\expandafter\def\csname cnt@rev.desc.d6.sspro.p2.C.tdiag.grid2-d3.boot-share.ci95.0\endcsname{1}
\expandafter\def\csname pct@rev.desc.d6.sspro.p2.C.tdiag.grid2-d3.boot-share.ci95.0\endcsname{54.2}
\expandafter\def\csname ppt@rev.desc.d6.sspro.p2.C.tdiag.grid2-d3.boot-share.ci95.0\endcsname{$+$54.2}
\expandafter\def\csname val@rev.desc.d6.sspro.p2.C.tdiag.grid2-d3.boot-share.ci95.1\endcsname{0.917}
\expandafter\def\csname cnt@rev.desc.d6.sspro.p2.C.tdiag.grid2-d3.boot-share.ci95.1\endcsname{1}
\expandafter\def\csname pct@rev.desc.d6.sspro.p2.C.tdiag.grid2-d3.boot-share.ci95.1\endcsname{91.7}
\expandafter\def\csname ppt@rev.desc.d6.sspro.p2.C.tdiag.grid2-d3.boot-share.ci95.1\endcsname{$+$91.7}
\expandafter\def\csname val@rev.desc.d6.sspro.p2.C.tdiag.grid2-d3.boot-share.n-apps\endcsname{6}
\expandafter\def\csname cnt@rev.desc.d6.sspro.p2.C.tdiag.grid2-d3.boot-share.n-apps\endcsname{6}
\expandafter\def\csname pct@rev.desc.d6.sspro.p2.C.tdiag.grid2-d3.boot-share.n-apps\endcsname{600.0}
\expandafter\def\csname ppt@rev.desc.d6.sspro.p2.C.tdiag.grid2-d3.boot-share.n-apps\endcsname{$+$600.0}
\expandafter\def\csname val@rev.desc.d6.sspro.p2.C.tdiag.grid2-d3.boot-share.n-boot-valid\endcsname{4000}
\expandafter\def\csname cnt@rev.desc.d6.sspro.p2.C.tdiag.grid2-d3.boot-share.n-boot-valid\endcsname{4{,}000}
\expandafter\def\csname pct@rev.desc.d6.sspro.p2.C.tdiag.grid2-d3.boot-share.n-boot-valid\endcsname{400000.0}
\expandafter\def\csname ppt@rev.desc.d6.sspro.p2.C.tdiag.grid2-d3.boot-share.n-boot-valid\endcsname{$+$400000.0}
\expandafter\def\csname val@rev.desc.d6.sspro.p2.C.tdiag.grid2-d3.boot-share.value\endcsname{0.771}
\expandafter\def\csname cnt@rev.desc.d6.sspro.p2.C.tdiag.grid2-d3.boot-share.value\endcsname{1}
\expandafter\def\csname pct@rev.desc.d6.sspro.p2.C.tdiag.grid2-d3.boot-share.value\endcsname{77.1}
\expandafter\def\csname ppt@rev.desc.d6.sspro.p2.C.tdiag.grid2-d3.boot-share.value\endcsname{$+$77.1}
\expandafter\def\csname val@rev.desc.d6.sspro.p2.C.tdiag.grid2-d3.n\endcsname{48}
\expandafter\def\csname cnt@rev.desc.d6.sspro.p2.C.tdiag.grid2-d3.n\endcsname{48}
\expandafter\def\csname pct@rev.desc.d6.sspro.p2.C.tdiag.grid2-d3.n\endcsname{4800.0}
\expandafter\def\csname ppt@rev.desc.d6.sspro.p2.C.tdiag.grid2-d3.n\endcsname{$+$4800.0}
\expandafter\def\csname val@rev.desc.d6.sspro.p2.C.tdiag.grid2-d3.resid-agree-stable\endcsname{-0.027}
\expandafter\def\csname cnt@rev.desc.d6.sspro.p2.C.tdiag.grid2-d3.resid-agree-stable\endcsname{0}
\expandafter\def\csname pct@rev.desc.d6.sspro.p2.C.tdiag.grid2-d3.resid-agree-stable\endcsname{$-$2.7}
\expandafter\def\csname ppt@rev.desc.d6.sspro.p2.C.tdiag.grid2-d3.resid-agree-stable\endcsname{$-$2.7}
\expandafter\def\csname val@rev.desc.d6.sspro.p2.C.tdiag.grid2-d3.resid-disagree\endcsname{0.000}
\expandafter\def\csname cnt@rev.desc.d6.sspro.p2.C.tdiag.grid2-d3.resid-disagree\endcsname{0}
\expandafter\def\csname pct@rev.desc.d6.sspro.p2.C.tdiag.grid2-d3.resid-disagree\endcsname{0.0}
\expandafter\def\csname ppt@rev.desc.d6.sspro.p2.C.tdiag.grid2-d3.resid-disagree\endcsname{0.0}
\expandafter\def\csname val@rev.desc.d6.sspro.p2.C.tdiag.grid2-d3.share-agree-stable\endcsname{0.771}
\expandafter\def\csname cnt@rev.desc.d6.sspro.p2.C.tdiag.grid2-d3.share-agree-stable\endcsname{1}
\expandafter\def\csname pct@rev.desc.d6.sspro.p2.C.tdiag.grid2-d3.share-agree-stable\endcsname{77.1}
\expandafter\def\csname ppt@rev.desc.d6.sspro.p2.C.tdiag.grid2-d3.share-agree-stable\endcsname{$+$77.1}
\expandafter\def\csname val@rev.desc.d6.sspro.p2.C.tdiag.zoomclick-d3.N-agree-stable\endcsname{39}
\expandafter\def\csname cnt@rev.desc.d6.sspro.p2.C.tdiag.zoomclick-d3.N-agree-stable\endcsname{39}
\expandafter\def\csname pct@rev.desc.d6.sspro.p2.C.tdiag.zoomclick-d3.N-agree-stable\endcsname{3900.0}
\expandafter\def\csname ppt@rev.desc.d6.sspro.p2.C.tdiag.zoomclick-d3.N-agree-stable\endcsname{$+$3900.0}
\expandafter\def\csname val@rev.desc.d6.sspro.p2.C.tdiag.zoomclick-d3.N-disagree\endcsname{8}
\expandafter\def\csname cnt@rev.desc.d6.sspro.p2.C.tdiag.zoomclick-d3.N-disagree\endcsname{8}
\expandafter\def\csname pct@rev.desc.d6.sspro.p2.C.tdiag.zoomclick-d3.N-disagree\endcsname{800.0}
\expandafter\def\csname ppt@rev.desc.d6.sspro.p2.C.tdiag.zoomclick-d3.N-disagree\endcsname{$+$800.0}
\expandafter\def\csname val@rev.desc.d6.sspro.p2.C.tdiag.zoomclick-d3.boot-resid.ci95.0\endcsname{0.000}
\expandafter\def\csname cnt@rev.desc.d6.sspro.p2.C.tdiag.zoomclick-d3.boot-resid.ci95.0\endcsname{0}
\expandafter\def\csname pct@rev.desc.d6.sspro.p2.C.tdiag.zoomclick-d3.boot-resid.ci95.0\endcsname{0.0}
\expandafter\def\csname ppt@rev.desc.d6.sspro.p2.C.tdiag.zoomclick-d3.boot-resid.ci95.0\endcsname{0.0}
\expandafter\def\csname val@rev.desc.d6.sspro.p2.C.tdiag.zoomclick-d3.boot-resid.ci95.1\endcsname{0.000}
\expandafter\def\csname cnt@rev.desc.d6.sspro.p2.C.tdiag.zoomclick-d3.boot-resid.ci95.1\endcsname{0}
\expandafter\def\csname pct@rev.desc.d6.sspro.p2.C.tdiag.zoomclick-d3.boot-resid.ci95.1\endcsname{0.0}
\expandafter\def\csname ppt@rev.desc.d6.sspro.p2.C.tdiag.zoomclick-d3.boot-resid.ci95.1\endcsname{0.0}
\expandafter\def\csname val@rev.desc.d6.sspro.p2.C.tdiag.zoomclick-d3.boot-resid.n-apps\endcsname{6}
\expandafter\def\csname cnt@rev.desc.d6.sspro.p2.C.tdiag.zoomclick-d3.boot-resid.n-apps\endcsname{6}
\expandafter\def\csname pct@rev.desc.d6.sspro.p2.C.tdiag.zoomclick-d3.boot-resid.n-apps\endcsname{600.0}
\expandafter\def\csname ppt@rev.desc.d6.sspro.p2.C.tdiag.zoomclick-d3.boot-resid.n-apps\endcsname{$+$600.0}
\expandafter\def\csname val@rev.desc.d6.sspro.p2.C.tdiag.zoomclick-d3.boot-resid.n-boot-valid\endcsname{4000}
\expandafter\def\csname cnt@rev.desc.d6.sspro.p2.C.tdiag.zoomclick-d3.boot-resid.n-boot-valid\endcsname{4{,}000}
\expandafter\def\csname pct@rev.desc.d6.sspro.p2.C.tdiag.zoomclick-d3.boot-resid.n-boot-valid\endcsname{400000.0}
\expandafter\def\csname ppt@rev.desc.d6.sspro.p2.C.tdiag.zoomclick-d3.boot-resid.n-boot-valid\endcsname{$+$400000.0}
\expandafter\def\csname val@rev.desc.d6.sspro.p2.C.tdiag.zoomclick-d3.boot-resid.value\endcsname{0.000}
\expandafter\def\csname cnt@rev.desc.d6.sspro.p2.C.tdiag.zoomclick-d3.boot-resid.value\endcsname{0}
\expandafter\def\csname pct@rev.desc.d6.sspro.p2.C.tdiag.zoomclick-d3.boot-resid.value\endcsname{0.0}
\expandafter\def\csname ppt@rev.desc.d6.sspro.p2.C.tdiag.zoomclick-d3.boot-resid.value\endcsname{0.0}
\expandafter\def\csname val@rev.desc.d6.sspro.p2.C.tdiag.zoomclick-d3.boot-share.ci95.0\endcsname{0.583}
\expandafter\def\csname cnt@rev.desc.d6.sspro.p2.C.tdiag.zoomclick-d3.boot-share.ci95.0\endcsname{1}
\expandafter\def\csname pct@rev.desc.d6.sspro.p2.C.tdiag.zoomclick-d3.boot-share.ci95.0\endcsname{58.3}
\expandafter\def\csname ppt@rev.desc.d6.sspro.p2.C.tdiag.zoomclick-d3.boot-share.ci95.0\endcsname{$+$58.3}
\expandafter\def\csname val@rev.desc.d6.sspro.p2.C.tdiag.zoomclick-d3.boot-share.ci95.1\endcsname{0.958}
\expandafter\def\csname cnt@rev.desc.d6.sspro.p2.C.tdiag.zoomclick-d3.boot-share.ci95.1\endcsname{1}
\expandafter\def\csname pct@rev.desc.d6.sspro.p2.C.tdiag.zoomclick-d3.boot-share.ci95.1\endcsname{95.8}
\expandafter\def\csname ppt@rev.desc.d6.sspro.p2.C.tdiag.zoomclick-d3.boot-share.ci95.1\endcsname{$+$95.8}
\expandafter\def\csname val@rev.desc.d6.sspro.p2.C.tdiag.zoomclick-d3.boot-share.n-apps\endcsname{6}
\expandafter\def\csname cnt@rev.desc.d6.sspro.p2.C.tdiag.zoomclick-d3.boot-share.n-apps\endcsname{6}
\expandafter\def\csname pct@rev.desc.d6.sspro.p2.C.tdiag.zoomclick-d3.boot-share.n-apps\endcsname{600.0}
\expandafter\def\csname ppt@rev.desc.d6.sspro.p2.C.tdiag.zoomclick-d3.boot-share.n-apps\endcsname{$+$600.0}
\expandafter\def\csname val@rev.desc.d6.sspro.p2.C.tdiag.zoomclick-d3.boot-share.n-boot-valid\endcsname{4000}
\expandafter\def\csname cnt@rev.desc.d6.sspro.p2.C.tdiag.zoomclick-d3.boot-share.n-boot-valid\endcsname{4{,}000}
\expandafter\def\csname pct@rev.desc.d6.sspro.p2.C.tdiag.zoomclick-d3.boot-share.n-boot-valid\endcsname{400000.0}
\expandafter\def\csname ppt@rev.desc.d6.sspro.p2.C.tdiag.zoomclick-d3.boot-share.n-boot-valid\endcsname{$+$400000.0}
\expandafter\def\csname val@rev.desc.d6.sspro.p2.C.tdiag.zoomclick-d3.boot-share.value\endcsname{0.812}
\expandafter\def\csname cnt@rev.desc.d6.sspro.p2.C.tdiag.zoomclick-d3.boot-share.value\endcsname{1}
\expandafter\def\csname pct@rev.desc.d6.sspro.p2.C.tdiag.zoomclick-d3.boot-share.value\endcsname{81.2}
\expandafter\def\csname ppt@rev.desc.d6.sspro.p2.C.tdiag.zoomclick-d3.boot-share.value\endcsname{$+$81.2}
\expandafter\def\csname val@rev.desc.d6.sspro.p2.C.tdiag.zoomclick-d3.n\endcsname{48}
\expandafter\def\csname cnt@rev.desc.d6.sspro.p2.C.tdiag.zoomclick-d3.n\endcsname{48}
\expandafter\def\csname pct@rev.desc.d6.sspro.p2.C.tdiag.zoomclick-d3.n\endcsname{4800.0}
\expandafter\def\csname ppt@rev.desc.d6.sspro.p2.C.tdiag.zoomclick-d3.n\endcsname{$+$4800.0}
\expandafter\def\csname val@rev.desc.d6.sspro.p2.C.tdiag.zoomclick-d3.resid-agree-stable\endcsname{0.000}
\expandafter\def\csname cnt@rev.desc.d6.sspro.p2.C.tdiag.zoomclick-d3.resid-agree-stable\endcsname{0}
\expandafter\def\csname pct@rev.desc.d6.sspro.p2.C.tdiag.zoomclick-d3.resid-agree-stable\endcsname{0.0}
\expandafter\def\csname ppt@rev.desc.d6.sspro.p2.C.tdiag.zoomclick-d3.resid-agree-stable\endcsname{0.0}
\expandafter\def\csname val@rev.desc.d6.sspro.p2.C.tdiag.zoomclick-d3.resid-disagree\endcsname{0.250}
\expandafter\def\csname cnt@rev.desc.d6.sspro.p2.C.tdiag.zoomclick-d3.resid-disagree\endcsname{0}
\expandafter\def\csname pct@rev.desc.d6.sspro.p2.C.tdiag.zoomclick-d3.resid-disagree\endcsname{25.0}
\expandafter\def\csname ppt@rev.desc.d6.sspro.p2.C.tdiag.zoomclick-d3.resid-disagree\endcsname{$+$25.0}
\expandafter\def\csname val@rev.desc.d6.sspro.p2.C.tdiag.zoomclick-d3.share-agree-stable\endcsname{0.812}
\expandafter\def\csname cnt@rev.desc.d6.sspro.p2.C.tdiag.zoomclick-d3.share-agree-stable\endcsname{1}
\expandafter\def\csname pct@rev.desc.d6.sspro.p2.C.tdiag.zoomclick-d3.share-agree-stable\endcsname{81.2}
\expandafter\def\csname ppt@rev.desc.d6.sspro.p2.C.tdiag.zoomclick-d3.share-agree-stable\endcsname{$+$81.2}
\expandafter\def\csname val@rev.desc.d6.sspro.p2.T.box.catalog-pooled.N-agree-stable\endcsname{276}
\expandafter\def\csname cnt@rev.desc.d6.sspro.p2.T.box.catalog-pooled.N-agree-stable\endcsname{276}
\expandafter\def\csname pct@rev.desc.d6.sspro.p2.T.box.catalog-pooled.N-agree-stable\endcsname{27600.0}
\expandafter\def\csname ppt@rev.desc.d6.sspro.p2.T.box.catalog-pooled.N-agree-stable\endcsname{$+$27600.0}
\expandafter\def\csname val@rev.desc.d6.sspro.p2.T.box.catalog-pooled.N-disagree\endcsname{101}
\expandafter\def\csname cnt@rev.desc.d6.sspro.p2.T.box.catalog-pooled.N-disagree\endcsname{101}
\expandafter\def\csname pct@rev.desc.d6.sspro.p2.T.box.catalog-pooled.N-disagree\endcsname{10100.0}
\expandafter\def\csname ppt@rev.desc.d6.sspro.p2.T.box.catalog-pooled.N-disagree\endcsname{$+$10100.0}
\expandafter\def\csname val@rev.desc.d6.sspro.p2.T.box.catalog-pooled.boot-resid.ci95.0\endcsname{-0.018}
\expandafter\def\csname cnt@rev.desc.d6.sspro.p2.T.box.catalog-pooled.boot-resid.ci95.0\endcsname{0}
\expandafter\def\csname pct@rev.desc.d6.sspro.p2.T.box.catalog-pooled.boot-resid.ci95.0\endcsname{$-$1.8}
\expandafter\def\csname ppt@rev.desc.d6.sspro.p2.T.box.catalog-pooled.boot-resid.ci95.0\endcsname{$-$1.8}
\expandafter\def\csname val@rev.desc.d6.sspro.p2.T.box.catalog-pooled.boot-resid.ci95.1\endcsname{0.000}
\expandafter\def\csname cnt@rev.desc.d6.sspro.p2.T.box.catalog-pooled.boot-resid.ci95.1\endcsname{0}
\expandafter\def\csname pct@rev.desc.d6.sspro.p2.T.box.catalog-pooled.boot-resid.ci95.1\endcsname{0.0}
\expandafter\def\csname ppt@rev.desc.d6.sspro.p2.T.box.catalog-pooled.boot-resid.ci95.1\endcsname{0.0}
\expandafter\def\csname val@rev.desc.d6.sspro.p2.T.box.catalog-pooled.boot-resid.n-apps\endcsname{12}
\expandafter\def\csname cnt@rev.desc.d6.sspro.p2.T.box.catalog-pooled.boot-resid.n-apps\endcsname{12}
\expandafter\def\csname pct@rev.desc.d6.sspro.p2.T.box.catalog-pooled.boot-resid.n-apps\endcsname{1200.0}
\expandafter\def\csname ppt@rev.desc.d6.sspro.p2.T.box.catalog-pooled.boot-resid.n-apps\endcsname{$+$1200.0}
\expandafter\def\csname val@rev.desc.d6.sspro.p2.T.box.catalog-pooled.boot-resid.n-boot-valid\endcsname{4000}
\expandafter\def\csname cnt@rev.desc.d6.sspro.p2.T.box.catalog-pooled.boot-resid.n-boot-valid\endcsname{4{,}000}
\expandafter\def\csname pct@rev.desc.d6.sspro.p2.T.box.catalog-pooled.boot-resid.n-boot-valid\endcsname{400000.0}
\expandafter\def\csname ppt@rev.desc.d6.sspro.p2.T.box.catalog-pooled.boot-resid.n-boot-valid\endcsname{$+$400000.0}
\expandafter\def\csname val@rev.desc.d6.sspro.p2.T.box.catalog-pooled.boot-resid.value\endcsname{-0.007}
\expandafter\def\csname cnt@rev.desc.d6.sspro.p2.T.box.catalog-pooled.boot-resid.value\endcsname{0}
\expandafter\def\csname pct@rev.desc.d6.sspro.p2.T.box.catalog-pooled.boot-resid.value\endcsname{$-$0.7}
\expandafter\def\csname ppt@rev.desc.d6.sspro.p2.T.box.catalog-pooled.boot-resid.value\endcsname{$-$0.7}
\expandafter\def\csname val@rev.desc.d6.sspro.p2.T.box.catalog-pooled.boot-share.ci95.0\endcsname{0.648}
\expandafter\def\csname cnt@rev.desc.d6.sspro.p2.T.box.catalog-pooled.boot-share.ci95.0\endcsname{1}
\expandafter\def\csname pct@rev.desc.d6.sspro.p2.T.box.catalog-pooled.boot-share.ci95.0\endcsname{64.8}
\expandafter\def\csname ppt@rev.desc.d6.sspro.p2.T.box.catalog-pooled.boot-share.ci95.0\endcsname{$+$64.8}
\expandafter\def\csname val@rev.desc.d6.sspro.p2.T.box.catalog-pooled.boot-share.ci95.1\endcsname{0.779}
\expandafter\def\csname cnt@rev.desc.d6.sspro.p2.T.box.catalog-pooled.boot-share.ci95.1\endcsname{1}
\expandafter\def\csname pct@rev.desc.d6.sspro.p2.T.box.catalog-pooled.boot-share.ci95.1\endcsname{77.9}
\expandafter\def\csname ppt@rev.desc.d6.sspro.p2.T.box.catalog-pooled.boot-share.ci95.1\endcsname{$+$77.9}
\expandafter\def\csname val@rev.desc.d6.sspro.p2.T.box.catalog-pooled.boot-share.n-apps\endcsname{12}
\expandafter\def\csname cnt@rev.desc.d6.sspro.p2.T.box.catalog-pooled.boot-share.n-apps\endcsname{12}
\expandafter\def\csname pct@rev.desc.d6.sspro.p2.T.box.catalog-pooled.boot-share.n-apps\endcsname{1200.0}
\expandafter\def\csname ppt@rev.desc.d6.sspro.p2.T.box.catalog-pooled.boot-share.n-apps\endcsname{$+$1200.0}
\expandafter\def\csname val@rev.desc.d6.sspro.p2.T.box.catalog-pooled.boot-share.n-boot-valid\endcsname{4000}
\expandafter\def\csname cnt@rev.desc.d6.sspro.p2.T.box.catalog-pooled.boot-share.n-boot-valid\endcsname{4{,}000}
\expandafter\def\csname pct@rev.desc.d6.sspro.p2.T.box.catalog-pooled.boot-share.n-boot-valid\endcsname{400000.0}
\expandafter\def\csname ppt@rev.desc.d6.sspro.p2.T.box.catalog-pooled.boot-share.n-boot-valid\endcsname{$+$400000.0}
\expandafter\def\csname val@rev.desc.d6.sspro.p2.T.box.catalog-pooled.boot-share.value\endcsname{0.719}
\expandafter\def\csname cnt@rev.desc.d6.sspro.p2.T.box.catalog-pooled.boot-share.value\endcsname{1}
\expandafter\def\csname pct@rev.desc.d6.sspro.p2.T.box.catalog-pooled.boot-share.value\endcsname{71.9}
\expandafter\def\csname ppt@rev.desc.d6.sspro.p2.T.box.catalog-pooled.boot-share.value\endcsname{$+$71.9}
\expandafter\def\csname val@rev.desc.d6.sspro.p2.T.box.catalog-pooled.gap\endcsname{0.096}
\expandafter\def\csname cnt@rev.desc.d6.sspro.p2.T.box.catalog-pooled.gap\endcsname{0}
\expandafter\def\csname pct@rev.desc.d6.sspro.p2.T.box.catalog-pooled.gap\endcsname{9.6}
\expandafter\def\csname ppt@rev.desc.d6.sspro.p2.T.box.catalog-pooled.gap\endcsname{$+$9.6}
\expandafter\def\csname val@rev.desc.d6.sspro.p2.T.box.catalog-pooled.n\endcsname{384}
\expandafter\def\csname cnt@rev.desc.d6.sspro.p2.T.box.catalog-pooled.n\endcsname{384}
\expandafter\def\csname pct@rev.desc.d6.sspro.p2.T.box.catalog-pooled.n\endcsname{38400.0}
\expandafter\def\csname ppt@rev.desc.d6.sspro.p2.T.box.catalog-pooled.n\endcsname{$+$38400.0}
\expandafter\def\csname val@rev.desc.d6.sspro.p2.T.box.catalog-pooled.resid-agree-stable\endcsname{-0.007}
\expandafter\def\csname cnt@rev.desc.d6.sspro.p2.T.box.catalog-pooled.resid-agree-stable\endcsname{0}
\expandafter\def\csname pct@rev.desc.d6.sspro.p2.T.box.catalog-pooled.resid-agree-stable\endcsname{$-$0.7}
\expandafter\def\csname ppt@rev.desc.d6.sspro.p2.T.box.catalog-pooled.resid-agree-stable\endcsname{$-$0.7}
\expandafter\def\csname val@rev.desc.d6.sspro.p2.T.box.catalog-pooled.resid-disagree\endcsname{0.089}
\expandafter\def\csname cnt@rev.desc.d6.sspro.p2.T.box.catalog-pooled.resid-disagree\endcsname{0}
\expandafter\def\csname pct@rev.desc.d6.sspro.p2.T.box.catalog-pooled.resid-disagree\endcsname{8.9}
\expandafter\def\csname ppt@rev.desc.d6.sspro.p2.T.box.catalog-pooled.resid-disagree\endcsname{$+$8.9}
\expandafter\def\csname val@rev.desc.d6.sspro.p2.T.box.catalog-pooled.share-agree-stable\endcsname{0.719}
\expandafter\def\csname cnt@rev.desc.d6.sspro.p2.T.box.catalog-pooled.share-agree-stable\endcsname{1}
\expandafter\def\csname pct@rev.desc.d6.sspro.p2.T.box.catalog-pooled.share-agree-stable\endcsname{71.9}
\expandafter\def\csname ppt@rev.desc.d6.sspro.p2.T.box.catalog-pooled.share-agree-stable\endcsname{$+$71.9}
\expandafter\def\csname val@rev.desc.d6.sspro.p2.T.box.center-d1.N-agree-stable\endcsname{70}
\expandafter\def\csname cnt@rev.desc.d6.sspro.p2.T.box.center-d1.N-agree-stable\endcsname{70}
\expandafter\def\csname pct@rev.desc.d6.sspro.p2.T.box.center-d1.N-agree-stable\endcsname{7000.0}
\expandafter\def\csname ppt@rev.desc.d6.sspro.p2.T.box.center-d1.N-agree-stable\endcsname{$+$7000.0}
\expandafter\def\csname val@rev.desc.d6.sspro.p2.T.box.center-d1.N-disagree\endcsname{25}
\expandafter\def\csname cnt@rev.desc.d6.sspro.p2.T.box.center-d1.N-disagree\endcsname{25}
\expandafter\def\csname pct@rev.desc.d6.sspro.p2.T.box.center-d1.N-disagree\endcsname{2500.0}
\expandafter\def\csname ppt@rev.desc.d6.sspro.p2.T.box.center-d1.N-disagree\endcsname{$+$2500.0}
\expandafter\def\csname val@rev.desc.d6.sspro.p2.T.box.center-d1.boot-resid.ci95.0\endcsname{0.000}
\expandafter\def\csname cnt@rev.desc.d6.sspro.p2.T.box.center-d1.boot-resid.ci95.0\endcsname{0}
\expandafter\def\csname pct@rev.desc.d6.sspro.p2.T.box.center-d1.boot-resid.ci95.0\endcsname{0.0}
\expandafter\def\csname ppt@rev.desc.d6.sspro.p2.T.box.center-d1.boot-resid.ci95.0\endcsname{0.0}
\expandafter\def\csname val@rev.desc.d6.sspro.p2.T.box.center-d1.boot-resid.ci95.1\endcsname{0.000}
\expandafter\def\csname cnt@rev.desc.d6.sspro.p2.T.box.center-d1.boot-resid.ci95.1\endcsname{0}
\expandafter\def\csname pct@rev.desc.d6.sspro.p2.T.box.center-d1.boot-resid.ci95.1\endcsname{0.0}
\expandafter\def\csname ppt@rev.desc.d6.sspro.p2.T.box.center-d1.boot-resid.ci95.1\endcsname{0.0}
\expandafter\def\csname val@rev.desc.d6.sspro.p2.T.box.center-d1.boot-resid.n-apps\endcsname{12}
\expandafter\def\csname cnt@rev.desc.d6.sspro.p2.T.box.center-d1.boot-resid.n-apps\endcsname{12}
\expandafter\def\csname pct@rev.desc.d6.sspro.p2.T.box.center-d1.boot-resid.n-apps\endcsname{1200.0}
\expandafter\def\csname ppt@rev.desc.d6.sspro.p2.T.box.center-d1.boot-resid.n-apps\endcsname{$+$1200.0}
\expandafter\def\csname val@rev.desc.d6.sspro.p2.T.box.center-d1.boot-resid.n-boot-valid\endcsname{4000}
\expandafter\def\csname cnt@rev.desc.d6.sspro.p2.T.box.center-d1.boot-resid.n-boot-valid\endcsname{4{,}000}
\expandafter\def\csname pct@rev.desc.d6.sspro.p2.T.box.center-d1.boot-resid.n-boot-valid\endcsname{400000.0}
\expandafter\def\csname ppt@rev.desc.d6.sspro.p2.T.box.center-d1.boot-resid.n-boot-valid\endcsname{$+$400000.0}
\expandafter\def\csname val@rev.desc.d6.sspro.p2.T.box.center-d1.boot-resid.value\endcsname{0.000}
\expandafter\def\csname cnt@rev.desc.d6.sspro.p2.T.box.center-d1.boot-resid.value\endcsname{0}
\expandafter\def\csname pct@rev.desc.d6.sspro.p2.T.box.center-d1.boot-resid.value\endcsname{0.0}
\expandafter\def\csname ppt@rev.desc.d6.sspro.p2.T.box.center-d1.boot-resid.value\endcsname{0.0}
\expandafter\def\csname val@rev.desc.d6.sspro.p2.T.box.center-d1.boot-share.ci95.0\endcsname{0.656}
\expandafter\def\csname cnt@rev.desc.d6.sspro.p2.T.box.center-d1.boot-share.ci95.0\endcsname{1}
\expandafter\def\csname pct@rev.desc.d6.sspro.p2.T.box.center-d1.boot-share.ci95.0\endcsname{65.6}
\expandafter\def\csname ppt@rev.desc.d6.sspro.p2.T.box.center-d1.boot-share.ci95.0\endcsname{$+$65.6}
\expandafter\def\csname val@rev.desc.d6.sspro.p2.T.box.center-d1.boot-share.ci95.1\endcsname{0.802}
\expandafter\def\csname cnt@rev.desc.d6.sspro.p2.T.box.center-d1.boot-share.ci95.1\endcsname{1}
\expandafter\def\csname pct@rev.desc.d6.sspro.p2.T.box.center-d1.boot-share.ci95.1\endcsname{80.2}
\expandafter\def\csname ppt@rev.desc.d6.sspro.p2.T.box.center-d1.boot-share.ci95.1\endcsname{$+$80.2}
\expandafter\def\csname val@rev.desc.d6.sspro.p2.T.box.center-d1.boot-share.n-apps\endcsname{12}
\expandafter\def\csname cnt@rev.desc.d6.sspro.p2.T.box.center-d1.boot-share.n-apps\endcsname{12}
\expandafter\def\csname pct@rev.desc.d6.sspro.p2.T.box.center-d1.boot-share.n-apps\endcsname{1200.0}
\expandafter\def\csname ppt@rev.desc.d6.sspro.p2.T.box.center-d1.boot-share.n-apps\endcsname{$+$1200.0}
\expandafter\def\csname val@rev.desc.d6.sspro.p2.T.box.center-d1.boot-share.n-boot-valid\endcsname{4000}
\expandafter\def\csname cnt@rev.desc.d6.sspro.p2.T.box.center-d1.boot-share.n-boot-valid\endcsname{4{,}000}
\expandafter\def\csname pct@rev.desc.d6.sspro.p2.T.box.center-d1.boot-share.n-boot-valid\endcsname{400000.0}
\expandafter\def\csname ppt@rev.desc.d6.sspro.p2.T.box.center-d1.boot-share.n-boot-valid\endcsname{$+$400000.0}
\expandafter\def\csname val@rev.desc.d6.sspro.p2.T.box.center-d1.boot-share.value\endcsname{0.729}
\expandafter\def\csname cnt@rev.desc.d6.sspro.p2.T.box.center-d1.boot-share.value\endcsname{1}
\expandafter\def\csname pct@rev.desc.d6.sspro.p2.T.box.center-d1.boot-share.value\endcsname{72.9}
\expandafter\def\csname ppt@rev.desc.d6.sspro.p2.T.box.center-d1.boot-share.value\endcsname{$+$72.9}
\expandafter\def\csname val@rev.desc.d6.sspro.p2.T.box.center-d1.n\endcsname{96}
\expandafter\def\csname cnt@rev.desc.d6.sspro.p2.T.box.center-d1.n\endcsname{96}
\expandafter\def\csname pct@rev.desc.d6.sspro.p2.T.box.center-d1.n\endcsname{9600.0}
\expandafter\def\csname ppt@rev.desc.d6.sspro.p2.T.box.center-d1.n\endcsname{$+$9600.0}
\expandafter\def\csname val@rev.desc.d6.sspro.p2.T.box.center-d1.resid-agree-stable\endcsname{0.000}
\expandafter\def\csname cnt@rev.desc.d6.sspro.p2.T.box.center-d1.resid-agree-stable\endcsname{0}
\expandafter\def\csname pct@rev.desc.d6.sspro.p2.T.box.center-d1.resid-agree-stable\endcsname{0.0}
\expandafter\def\csname ppt@rev.desc.d6.sspro.p2.T.box.center-d1.resid-agree-stable\endcsname{0.0}
\expandafter\def\csname val@rev.desc.d6.sspro.p2.T.box.center-d1.resid-disagree\endcsname{0.120}
\expandafter\def\csname cnt@rev.desc.d6.sspro.p2.T.box.center-d1.resid-disagree\endcsname{0}
\expandafter\def\csname pct@rev.desc.d6.sspro.p2.T.box.center-d1.resid-disagree\endcsname{12.0}
\expandafter\def\csname ppt@rev.desc.d6.sspro.p2.T.box.center-d1.resid-disagree\endcsname{$+$12.0}
\expandafter\def\csname val@rev.desc.d6.sspro.p2.T.box.center-d1.share-agree-stable\endcsname{0.729}
\expandafter\def\csname cnt@rev.desc.d6.sspro.p2.T.box.center-d1.share-agree-stable\endcsname{1}
\expandafter\def\csname pct@rev.desc.d6.sspro.p2.T.box.center-d1.share-agree-stable\endcsname{72.9}
\expandafter\def\csname ppt@rev.desc.d6.sspro.p2.T.box.center-d1.share-agree-stable\endcsname{$+$72.9}
\expandafter\def\csname val@rev.desc.d6.sspro.p2.T.box.context-d1.N-agree-stable\endcsname{70}
\expandafter\def\csname cnt@rev.desc.d6.sspro.p2.T.box.context-d1.N-agree-stable\endcsname{70}
\expandafter\def\csname pct@rev.desc.d6.sspro.p2.T.box.context-d1.N-agree-stable\endcsname{7000.0}
\expandafter\def\csname ppt@rev.desc.d6.sspro.p2.T.box.context-d1.N-agree-stable\endcsname{$+$7000.0}
\expandafter\def\csname val@rev.desc.d6.sspro.p2.T.box.context-d1.N-disagree\endcsname{25}
\expandafter\def\csname cnt@rev.desc.d6.sspro.p2.T.box.context-d1.N-disagree\endcsname{25}
\expandafter\def\csname pct@rev.desc.d6.sspro.p2.T.box.context-d1.N-disagree\endcsname{2500.0}
\expandafter\def\csname ppt@rev.desc.d6.sspro.p2.T.box.context-d1.N-disagree\endcsname{$+$2500.0}
\expandafter\def\csname val@rev.desc.d6.sspro.p2.T.box.context-d1.boot-resid.ci95.0\endcsname{-0.044}
\expandafter\def\csname cnt@rev.desc.d6.sspro.p2.T.box.context-d1.boot-resid.ci95.0\endcsname{0}
\expandafter\def\csname pct@rev.desc.d6.sspro.p2.T.box.context-d1.boot-resid.ci95.0\endcsname{$-$4.4}
\expandafter\def\csname ppt@rev.desc.d6.sspro.p2.T.box.context-d1.boot-resid.ci95.0\endcsname{$-$4.4}
\expandafter\def\csname val@rev.desc.d6.sspro.p2.T.box.context-d1.boot-resid.ci95.1\endcsname{0.000}
\expandafter\def\csname cnt@rev.desc.d6.sspro.p2.T.box.context-d1.boot-resid.ci95.1\endcsname{0}
\expandafter\def\csname pct@rev.desc.d6.sspro.p2.T.box.context-d1.boot-resid.ci95.1\endcsname{0.0}
\expandafter\def\csname ppt@rev.desc.d6.sspro.p2.T.box.context-d1.boot-resid.ci95.1\endcsname{0.0}
\expandafter\def\csname val@rev.desc.d6.sspro.p2.T.box.context-d1.boot-resid.n-apps\endcsname{12}
\expandafter\def\csname cnt@rev.desc.d6.sspro.p2.T.box.context-d1.boot-resid.n-apps\endcsname{12}
\expandafter\def\csname pct@rev.desc.d6.sspro.p2.T.box.context-d1.boot-resid.n-apps\endcsname{1200.0}
\expandafter\def\csname ppt@rev.desc.d6.sspro.p2.T.box.context-d1.boot-resid.n-apps\endcsname{$+$1200.0}
\expandafter\def\csname val@rev.desc.d6.sspro.p2.T.box.context-d1.boot-resid.n-boot-valid\endcsname{4000}
\expandafter\def\csname cnt@rev.desc.d6.sspro.p2.T.box.context-d1.boot-resid.n-boot-valid\endcsname{4{,}000}
\expandafter\def\csname pct@rev.desc.d6.sspro.p2.T.box.context-d1.boot-resid.n-boot-valid\endcsname{400000.0}
\expandafter\def\csname ppt@rev.desc.d6.sspro.p2.T.box.context-d1.boot-resid.n-boot-valid\endcsname{$+$400000.0}
\expandafter\def\csname val@rev.desc.d6.sspro.p2.T.box.context-d1.boot-resid.value\endcsname{-0.014}
\expandafter\def\csname cnt@rev.desc.d6.sspro.p2.T.box.context-d1.boot-resid.value\endcsname{0}
\expandafter\def\csname pct@rev.desc.d6.sspro.p2.T.box.context-d1.boot-resid.value\endcsname{$-$1.4}
\expandafter\def\csname ppt@rev.desc.d6.sspro.p2.T.box.context-d1.boot-resid.value\endcsname{$-$1.4}
\expandafter\def\csname val@rev.desc.d6.sspro.p2.T.box.context-d1.boot-share.ci95.0\endcsname{0.656}
\expandafter\def\csname cnt@rev.desc.d6.sspro.p2.T.box.context-d1.boot-share.ci95.0\endcsname{1}
\expandafter\def\csname pct@rev.desc.d6.sspro.p2.T.box.context-d1.boot-share.ci95.0\endcsname{65.6}
\expandafter\def\csname ppt@rev.desc.d6.sspro.p2.T.box.context-d1.boot-share.ci95.0\endcsname{$+$65.6}
\expandafter\def\csname val@rev.desc.d6.sspro.p2.T.box.context-d1.boot-share.ci95.1\endcsname{0.792}
\expandafter\def\csname cnt@rev.desc.d6.sspro.p2.T.box.context-d1.boot-share.ci95.1\endcsname{1}
\expandafter\def\csname pct@rev.desc.d6.sspro.p2.T.box.context-d1.boot-share.ci95.1\endcsname{79.2}
\expandafter\def\csname ppt@rev.desc.d6.sspro.p2.T.box.context-d1.boot-share.ci95.1\endcsname{$+$79.2}
\expandafter\def\csname val@rev.desc.d6.sspro.p2.T.box.context-d1.boot-share.n-apps\endcsname{12}
\expandafter\def\csname cnt@rev.desc.d6.sspro.p2.T.box.context-d1.boot-share.n-apps\endcsname{12}
\expandafter\def\csname pct@rev.desc.d6.sspro.p2.T.box.context-d1.boot-share.n-apps\endcsname{1200.0}
\expandafter\def\csname ppt@rev.desc.d6.sspro.p2.T.box.context-d1.boot-share.n-apps\endcsname{$+$1200.0}
\expandafter\def\csname val@rev.desc.d6.sspro.p2.T.box.context-d1.boot-share.n-boot-valid\endcsname{4000}
\expandafter\def\csname cnt@rev.desc.d6.sspro.p2.T.box.context-d1.boot-share.n-boot-valid\endcsname{4{,}000}
\expandafter\def\csname pct@rev.desc.d6.sspro.p2.T.box.context-d1.boot-share.n-boot-valid\endcsname{400000.0}
\expandafter\def\csname ppt@rev.desc.d6.sspro.p2.T.box.context-d1.boot-share.n-boot-valid\endcsname{$+$400000.0}
\expandafter\def\csname val@rev.desc.d6.sspro.p2.T.box.context-d1.boot-share.value\endcsname{0.729}
\expandafter\def\csname cnt@rev.desc.d6.sspro.p2.T.box.context-d1.boot-share.value\endcsname{1}
\expandafter\def\csname pct@rev.desc.d6.sspro.p2.T.box.context-d1.boot-share.value\endcsname{72.9}
\expandafter\def\csname ppt@rev.desc.d6.sspro.p2.T.box.context-d1.boot-share.value\endcsname{$+$72.9}
\expandafter\def\csname val@rev.desc.d6.sspro.p2.T.box.context-d1.n\endcsname{96}
\expandafter\def\csname cnt@rev.desc.d6.sspro.p2.T.box.context-d1.n\endcsname{96}
\expandafter\def\csname pct@rev.desc.d6.sspro.p2.T.box.context-d1.n\endcsname{9600.0}
\expandafter\def\csname ppt@rev.desc.d6.sspro.p2.T.box.context-d1.n\endcsname{$+$9600.0}
\expandafter\def\csname val@rev.desc.d6.sspro.p2.T.box.context-d1.resid-agree-stable\endcsname{-0.014}
\expandafter\def\csname cnt@rev.desc.d6.sspro.p2.T.box.context-d1.resid-agree-stable\endcsname{0}
\expandafter\def\csname pct@rev.desc.d6.sspro.p2.T.box.context-d1.resid-agree-stable\endcsname{$-$1.4}
\expandafter\def\csname ppt@rev.desc.d6.sspro.p2.T.box.context-d1.resid-agree-stable\endcsname{$-$1.4}
\expandafter\def\csname val@rev.desc.d6.sspro.p2.T.box.context-d1.resid-disagree\endcsname{-0.040}
\expandafter\def\csname cnt@rev.desc.d6.sspro.p2.T.box.context-d1.resid-disagree\endcsname{0}
\expandafter\def\csname pct@rev.desc.d6.sspro.p2.T.box.context-d1.resid-disagree\endcsname{$-$4.0}
\expandafter\def\csname ppt@rev.desc.d6.sspro.p2.T.box.context-d1.resid-disagree\endcsname{$-$4.0}
\expandafter\def\csname val@rev.desc.d6.sspro.p2.T.box.context-d1.share-agree-stable\endcsname{0.729}
\expandafter\def\csname cnt@rev.desc.d6.sspro.p2.T.box.context-d1.share-agree-stable\endcsname{1}
\expandafter\def\csname pct@rev.desc.d6.sspro.p2.T.box.context-d1.share-agree-stable\endcsname{72.9}
\expandafter\def\csname ppt@rev.desc.d6.sspro.p2.T.box.context-d1.share-agree-stable\endcsname{$+$72.9}
\expandafter\def\csname val@rev.desc.d6.sspro.p2.T.box.dual-d2.N-agree-stable\endcsname{69}
\expandafter\def\csname cnt@rev.desc.d6.sspro.p2.T.box.dual-d2.N-agree-stable\endcsname{69}
\expandafter\def\csname pct@rev.desc.d6.sspro.p2.T.box.dual-d2.N-agree-stable\endcsname{6900.0}
\expandafter\def\csname ppt@rev.desc.d6.sspro.p2.T.box.dual-d2.N-agree-stable\endcsname{$+$6900.0}
\expandafter\def\csname val@rev.desc.d6.sspro.p2.T.box.dual-d2.N-disagree\endcsname{26}
\expandafter\def\csname cnt@rev.desc.d6.sspro.p2.T.box.dual-d2.N-disagree\endcsname{26}
\expandafter\def\csname pct@rev.desc.d6.sspro.p2.T.box.dual-d2.N-disagree\endcsname{2600.0}
\expandafter\def\csname ppt@rev.desc.d6.sspro.p2.T.box.dual-d2.N-disagree\endcsname{$+$2600.0}
\expandafter\def\csname val@rev.desc.d6.sspro.p2.T.box.dual-d2.boot-resid.ci95.0\endcsname{0.000}
\expandafter\def\csname cnt@rev.desc.d6.sspro.p2.T.box.dual-d2.boot-resid.ci95.0\endcsname{0}
\expandafter\def\csname pct@rev.desc.d6.sspro.p2.T.box.dual-d2.boot-resid.ci95.0\endcsname{0.0}
\expandafter\def\csname ppt@rev.desc.d6.sspro.p2.T.box.dual-d2.boot-resid.ci95.0\endcsname{0.0}
\expandafter\def\csname val@rev.desc.d6.sspro.p2.T.box.dual-d2.boot-resid.ci95.1\endcsname{0.000}
\expandafter\def\csname cnt@rev.desc.d6.sspro.p2.T.box.dual-d2.boot-resid.ci95.1\endcsname{0}
\expandafter\def\csname pct@rev.desc.d6.sspro.p2.T.box.dual-d2.boot-resid.ci95.1\endcsname{0.0}
\expandafter\def\csname ppt@rev.desc.d6.sspro.p2.T.box.dual-d2.boot-resid.ci95.1\endcsname{0.0}
\expandafter\def\csname val@rev.desc.d6.sspro.p2.T.box.dual-d2.boot-resid.n-apps\endcsname{12}
\expandafter\def\csname cnt@rev.desc.d6.sspro.p2.T.box.dual-d2.boot-resid.n-apps\endcsname{12}
\expandafter\def\csname pct@rev.desc.d6.sspro.p2.T.box.dual-d2.boot-resid.n-apps\endcsname{1200.0}
\expandafter\def\csname ppt@rev.desc.d6.sspro.p2.T.box.dual-d2.boot-resid.n-apps\endcsname{$+$1200.0}
\expandafter\def\csname val@rev.desc.d6.sspro.p2.T.box.dual-d2.boot-resid.n-boot-valid\endcsname{4000}
\expandafter\def\csname cnt@rev.desc.d6.sspro.p2.T.box.dual-d2.boot-resid.n-boot-valid\endcsname{4{,}000}
\expandafter\def\csname pct@rev.desc.d6.sspro.p2.T.box.dual-d2.boot-resid.n-boot-valid\endcsname{400000.0}
\expandafter\def\csname ppt@rev.desc.d6.sspro.p2.T.box.dual-d2.boot-resid.n-boot-valid\endcsname{$+$400000.0}
\expandafter\def\csname val@rev.desc.d6.sspro.p2.T.box.dual-d2.boot-resid.value\endcsname{0.000}
\expandafter\def\csname cnt@rev.desc.d6.sspro.p2.T.box.dual-d2.boot-resid.value\endcsname{0}
\expandafter\def\csname pct@rev.desc.d6.sspro.p2.T.box.dual-d2.boot-resid.value\endcsname{0.0}
\expandafter\def\csname ppt@rev.desc.d6.sspro.p2.T.box.dual-d2.boot-resid.value\endcsname{0.0}
\expandafter\def\csname val@rev.desc.d6.sspro.p2.T.box.dual-d2.boot-share.ci95.0\endcsname{0.656}
\expandafter\def\csname cnt@rev.desc.d6.sspro.p2.T.box.dual-d2.boot-share.ci95.0\endcsname{1}
\expandafter\def\csname pct@rev.desc.d6.sspro.p2.T.box.dual-d2.boot-share.ci95.0\endcsname{65.6}
\expandafter\def\csname ppt@rev.desc.d6.sspro.p2.T.box.dual-d2.boot-share.ci95.0\endcsname{$+$65.6}
\expandafter\def\csname val@rev.desc.d6.sspro.p2.T.box.dual-d2.boot-share.ci95.1\endcsname{0.781}
\expandafter\def\csname cnt@rev.desc.d6.sspro.p2.T.box.dual-d2.boot-share.ci95.1\endcsname{1}
\expandafter\def\csname pct@rev.desc.d6.sspro.p2.T.box.dual-d2.boot-share.ci95.1\endcsname{78.1}
\expandafter\def\csname ppt@rev.desc.d6.sspro.p2.T.box.dual-d2.boot-share.ci95.1\endcsname{$+$78.1}
\expandafter\def\csname val@rev.desc.d6.sspro.p2.T.box.dual-d2.boot-share.n-apps\endcsname{12}
\expandafter\def\csname cnt@rev.desc.d6.sspro.p2.T.box.dual-d2.boot-share.n-apps\endcsname{12}
\expandafter\def\csname pct@rev.desc.d6.sspro.p2.T.box.dual-d2.boot-share.n-apps\endcsname{1200.0}
\expandafter\def\csname ppt@rev.desc.d6.sspro.p2.T.box.dual-d2.boot-share.n-apps\endcsname{$+$1200.0}
\expandafter\def\csname val@rev.desc.d6.sspro.p2.T.box.dual-d2.boot-share.n-boot-valid\endcsname{4000}
\expandafter\def\csname cnt@rev.desc.d6.sspro.p2.T.box.dual-d2.boot-share.n-boot-valid\endcsname{4{,}000}
\expandafter\def\csname pct@rev.desc.d6.sspro.p2.T.box.dual-d2.boot-share.n-boot-valid\endcsname{400000.0}
\expandafter\def\csname ppt@rev.desc.d6.sspro.p2.T.box.dual-d2.boot-share.n-boot-valid\endcsname{$+$400000.0}
\expandafter\def\csname val@rev.desc.d6.sspro.p2.T.box.dual-d2.boot-share.value\endcsname{0.719}
\expandafter\def\csname cnt@rev.desc.d6.sspro.p2.T.box.dual-d2.boot-share.value\endcsname{1}
\expandafter\def\csname pct@rev.desc.d6.sspro.p2.T.box.dual-d2.boot-share.value\endcsname{71.9}
\expandafter\def\csname ppt@rev.desc.d6.sspro.p2.T.box.dual-d2.boot-share.value\endcsname{$+$71.9}
\expandafter\def\csname val@rev.desc.d6.sspro.p2.T.box.dual-d2.n\endcsname{96}
\expandafter\def\csname cnt@rev.desc.d6.sspro.p2.T.box.dual-d2.n\endcsname{96}
\expandafter\def\csname pct@rev.desc.d6.sspro.p2.T.box.dual-d2.n\endcsname{9600.0}
\expandafter\def\csname ppt@rev.desc.d6.sspro.p2.T.box.dual-d2.n\endcsname{$+$9600.0}
\expandafter\def\csname val@rev.desc.d6.sspro.p2.T.box.dual-d2.resid-agree-stable\endcsname{0.000}
\expandafter\def\csname cnt@rev.desc.d6.sspro.p2.T.box.dual-d2.resid-agree-stable\endcsname{0}
\expandafter\def\csname pct@rev.desc.d6.sspro.p2.T.box.dual-d2.resid-agree-stable\endcsname{0.0}
\expandafter\def\csname ppt@rev.desc.d6.sspro.p2.T.box.dual-d2.resid-agree-stable\endcsname{0.0}
\expandafter\def\csname val@rev.desc.d6.sspro.p2.T.box.dual-d2.resid-disagree\endcsname{0.192}
\expandafter\def\csname cnt@rev.desc.d6.sspro.p2.T.box.dual-d2.resid-disagree\endcsname{0}
\expandafter\def\csname pct@rev.desc.d6.sspro.p2.T.box.dual-d2.resid-disagree\endcsname{19.2}
\expandafter\def\csname ppt@rev.desc.d6.sspro.p2.T.box.dual-d2.resid-disagree\endcsname{$+$19.2}
\expandafter\def\csname val@rev.desc.d6.sspro.p2.T.box.dual-d2.share-agree-stable\endcsname{0.719}
\expandafter\def\csname cnt@rev.desc.d6.sspro.p2.T.box.dual-d2.share-agree-stable\endcsname{1}
\expandafter\def\csname pct@rev.desc.d6.sspro.p2.T.box.dual-d2.share-agree-stable\endcsname{71.9}
\expandafter\def\csname ppt@rev.desc.d6.sspro.p2.T.box.dual-d2.share-agree-stable\endcsname{$+$71.9}
\expandafter\def\csname val@rev.desc.d6.sspro.p2.T.box.grid2-d3.N-agree-stable\endcsname{67}
\expandafter\def\csname cnt@rev.desc.d6.sspro.p2.T.box.grid2-d3.N-agree-stable\endcsname{67}
\expandafter\def\csname pct@rev.desc.d6.sspro.p2.T.box.grid2-d3.N-agree-stable\endcsname{6700.0}
\expandafter\def\csname ppt@rev.desc.d6.sspro.p2.T.box.grid2-d3.N-agree-stable\endcsname{$+$6700.0}
\expandafter\def\csname val@rev.desc.d6.sspro.p2.T.box.grid2-d3.N-disagree\endcsname{25}
\expandafter\def\csname cnt@rev.desc.d6.sspro.p2.T.box.grid2-d3.N-disagree\endcsname{25}
\expandafter\def\csname pct@rev.desc.d6.sspro.p2.T.box.grid2-d3.N-disagree\endcsname{2500.0}
\expandafter\def\csname ppt@rev.desc.d6.sspro.p2.T.box.grid2-d3.N-disagree\endcsname{$+$2500.0}
\expandafter\def\csname val@rev.desc.d6.sspro.p2.T.box.grid2-d3.boot-resid.ci95.0\endcsname{-0.045}
\expandafter\def\csname cnt@rev.desc.d6.sspro.p2.T.box.grid2-d3.boot-resid.ci95.0\endcsname{0}
\expandafter\def\csname pct@rev.desc.d6.sspro.p2.T.box.grid2-d3.boot-resid.ci95.0\endcsname{$-$4.5}
\expandafter\def\csname ppt@rev.desc.d6.sspro.p2.T.box.grid2-d3.boot-resid.ci95.0\endcsname{$-$4.5}
\expandafter\def\csname val@rev.desc.d6.sspro.p2.T.box.grid2-d3.boot-resid.ci95.1\endcsname{0.000}
\expandafter\def\csname cnt@rev.desc.d6.sspro.p2.T.box.grid2-d3.boot-resid.ci95.1\endcsname{0}
\expandafter\def\csname pct@rev.desc.d6.sspro.p2.T.box.grid2-d3.boot-resid.ci95.1\endcsname{0.0}
\expandafter\def\csname ppt@rev.desc.d6.sspro.p2.T.box.grid2-d3.boot-resid.ci95.1\endcsname{0.0}
\expandafter\def\csname val@rev.desc.d6.sspro.p2.T.box.grid2-d3.boot-resid.n-apps\endcsname{12}
\expandafter\def\csname cnt@rev.desc.d6.sspro.p2.T.box.grid2-d3.boot-resid.n-apps\endcsname{12}
\expandafter\def\csname pct@rev.desc.d6.sspro.p2.T.box.grid2-d3.boot-resid.n-apps\endcsname{1200.0}
\expandafter\def\csname ppt@rev.desc.d6.sspro.p2.T.box.grid2-d3.boot-resid.n-apps\endcsname{$+$1200.0}
\expandafter\def\csname val@rev.desc.d6.sspro.p2.T.box.grid2-d3.boot-resid.n-boot-valid\endcsname{4000}
\expandafter\def\csname cnt@rev.desc.d6.sspro.p2.T.box.grid2-d3.boot-resid.n-boot-valid\endcsname{4{,}000}
\expandafter\def\csname pct@rev.desc.d6.sspro.p2.T.box.grid2-d3.boot-resid.n-boot-valid\endcsname{400000.0}
\expandafter\def\csname ppt@rev.desc.d6.sspro.p2.T.box.grid2-d3.boot-resid.n-boot-valid\endcsname{$+$400000.0}
\expandafter\def\csname val@rev.desc.d6.sspro.p2.T.box.grid2-d3.boot-resid.value\endcsname{-0.015}
\expandafter\def\csname cnt@rev.desc.d6.sspro.p2.T.box.grid2-d3.boot-resid.value\endcsname{0}
\expandafter\def\csname pct@rev.desc.d6.sspro.p2.T.box.grid2-d3.boot-resid.value\endcsname{$-$1.5}
\expandafter\def\csname ppt@rev.desc.d6.sspro.p2.T.box.grid2-d3.boot-resid.value\endcsname{$-$1.5}
\expandafter\def\csname val@rev.desc.d6.sspro.p2.T.box.grid2-d3.boot-share.ci95.0\endcsname{0.615}
\expandafter\def\csname cnt@rev.desc.d6.sspro.p2.T.box.grid2-d3.boot-share.ci95.0\endcsname{1}
\expandafter\def\csname pct@rev.desc.d6.sspro.p2.T.box.grid2-d3.boot-share.ci95.0\endcsname{61.5}
\expandafter\def\csname ppt@rev.desc.d6.sspro.p2.T.box.grid2-d3.boot-share.ci95.0\endcsname{$+$61.5}
\expandafter\def\csname val@rev.desc.d6.sspro.p2.T.box.grid2-d3.boot-share.ci95.1\endcsname{0.771}
\expandafter\def\csname cnt@rev.desc.d6.sspro.p2.T.box.grid2-d3.boot-share.ci95.1\endcsname{1}
\expandafter\def\csname pct@rev.desc.d6.sspro.p2.T.box.grid2-d3.boot-share.ci95.1\endcsname{77.1}
\expandafter\def\csname ppt@rev.desc.d6.sspro.p2.T.box.grid2-d3.boot-share.ci95.1\endcsname{$+$77.1}
\expandafter\def\csname val@rev.desc.d6.sspro.p2.T.box.grid2-d3.boot-share.n-apps\endcsname{12}
\expandafter\def\csname cnt@rev.desc.d6.sspro.p2.T.box.grid2-d3.boot-share.n-apps\endcsname{12}
\expandafter\def\csname pct@rev.desc.d6.sspro.p2.T.box.grid2-d3.boot-share.n-apps\endcsname{1200.0}
\expandafter\def\csname ppt@rev.desc.d6.sspro.p2.T.box.grid2-d3.boot-share.n-apps\endcsname{$+$1200.0}
\expandafter\def\csname val@rev.desc.d6.sspro.p2.T.box.grid2-d3.boot-share.n-boot-valid\endcsname{4000}
\expandafter\def\csname cnt@rev.desc.d6.sspro.p2.T.box.grid2-d3.boot-share.n-boot-valid\endcsname{4{,}000}
\expandafter\def\csname pct@rev.desc.d6.sspro.p2.T.box.grid2-d3.boot-share.n-boot-valid\endcsname{400000.0}
\expandafter\def\csname ppt@rev.desc.d6.sspro.p2.T.box.grid2-d3.boot-share.n-boot-valid\endcsname{$+$400000.0}
\expandafter\def\csname val@rev.desc.d6.sspro.p2.T.box.grid2-d3.boot-share.value\endcsname{0.698}
\expandafter\def\csname cnt@rev.desc.d6.sspro.p2.T.box.grid2-d3.boot-share.value\endcsname{1}
\expandafter\def\csname pct@rev.desc.d6.sspro.p2.T.box.grid2-d3.boot-share.value\endcsname{69.8}
\expandafter\def\csname ppt@rev.desc.d6.sspro.p2.T.box.grid2-d3.boot-share.value\endcsname{$+$69.8}
\expandafter\def\csname val@rev.desc.d6.sspro.p2.T.box.grid2-d3.n\endcsname{96}
\expandafter\def\csname cnt@rev.desc.d6.sspro.p2.T.box.grid2-d3.n\endcsname{96}
\expandafter\def\csname pct@rev.desc.d6.sspro.p2.T.box.grid2-d3.n\endcsname{9600.0}
\expandafter\def\csname ppt@rev.desc.d6.sspro.p2.T.box.grid2-d3.n\endcsname{$+$9600.0}
\expandafter\def\csname val@rev.desc.d6.sspro.p2.T.box.grid2-d3.resid-agree-stable\endcsname{-0.015}
\expandafter\def\csname cnt@rev.desc.d6.sspro.p2.T.box.grid2-d3.resid-agree-stable\endcsname{0}
\expandafter\def\csname pct@rev.desc.d6.sspro.p2.T.box.grid2-d3.resid-agree-stable\endcsname{$-$1.5}
\expandafter\def\csname ppt@rev.desc.d6.sspro.p2.T.box.grid2-d3.resid-agree-stable\endcsname{$-$1.5}
\expandafter\def\csname val@rev.desc.d6.sspro.p2.T.box.grid2-d3.resid-disagree\endcsname{0.080}
\expandafter\def\csname cnt@rev.desc.d6.sspro.p2.T.box.grid2-d3.resid-disagree\endcsname{0}
\expandafter\def\csname pct@rev.desc.d6.sspro.p2.T.box.grid2-d3.resid-disagree\endcsname{8.0}
\expandafter\def\csname ppt@rev.desc.d6.sspro.p2.T.box.grid2-d3.resid-disagree\endcsname{$+$8.0}
\expandafter\def\csname val@rev.desc.d6.sspro.p2.T.box.grid2-d3.share-agree-stable\endcsname{0.698}
\expandafter\def\csname cnt@rev.desc.d6.sspro.p2.T.box.grid2-d3.share-agree-stable\endcsname{1}
\expandafter\def\csname pct@rev.desc.d6.sspro.p2.T.box.grid2-d3.share-agree-stable\endcsname{69.8}
\expandafter\def\csname ppt@rev.desc.d6.sspro.p2.T.box.grid2-d3.share-agree-stable\endcsname{$+$69.8}
\expandafter\def\csname val@rev.desc.d6.sspro.p2.T.box.zoomclick-d3.N-agree-stable\endcsname{71}
\expandafter\def\csname cnt@rev.desc.d6.sspro.p2.T.box.zoomclick-d3.N-agree-stable\endcsname{71}
\expandafter\def\csname pct@rev.desc.d6.sspro.p2.T.box.zoomclick-d3.N-agree-stable\endcsname{7100.0}
\expandafter\def\csname ppt@rev.desc.d6.sspro.p2.T.box.zoomclick-d3.N-agree-stable\endcsname{$+$7100.0}
\expandafter\def\csname val@rev.desc.d6.sspro.p2.T.box.zoomclick-d3.N-disagree\endcsname{24}
\expandafter\def\csname cnt@rev.desc.d6.sspro.p2.T.box.zoomclick-d3.N-disagree\endcsname{24}
\expandafter\def\csname pct@rev.desc.d6.sspro.p2.T.box.zoomclick-d3.N-disagree\endcsname{2400.0}
\expandafter\def\csname ppt@rev.desc.d6.sspro.p2.T.box.zoomclick-d3.N-disagree\endcsname{$+$2400.0}
\expandafter\def\csname val@rev.desc.d6.sspro.p2.T.box.zoomclick-d3.boot-resid.ci95.0\endcsname{0.000}
\expandafter\def\csname cnt@rev.desc.d6.sspro.p2.T.box.zoomclick-d3.boot-resid.ci95.0\endcsname{0}
\expandafter\def\csname pct@rev.desc.d6.sspro.p2.T.box.zoomclick-d3.boot-resid.ci95.0\endcsname{0.0}
\expandafter\def\csname ppt@rev.desc.d6.sspro.p2.T.box.zoomclick-d3.boot-resid.ci95.0\endcsname{0.0}
\expandafter\def\csname val@rev.desc.d6.sspro.p2.T.box.zoomclick-d3.boot-resid.ci95.1\endcsname{0.000}
\expandafter\def\csname cnt@rev.desc.d6.sspro.p2.T.box.zoomclick-d3.boot-resid.ci95.1\endcsname{0}
\expandafter\def\csname pct@rev.desc.d6.sspro.p2.T.box.zoomclick-d3.boot-resid.ci95.1\endcsname{0.0}
\expandafter\def\csname ppt@rev.desc.d6.sspro.p2.T.box.zoomclick-d3.boot-resid.ci95.1\endcsname{0.0}
\expandafter\def\csname val@rev.desc.d6.sspro.p2.T.box.zoomclick-d3.boot-resid.n-apps\endcsname{12}
\expandafter\def\csname cnt@rev.desc.d6.sspro.p2.T.box.zoomclick-d3.boot-resid.n-apps\endcsname{12}
\expandafter\def\csname pct@rev.desc.d6.sspro.p2.T.box.zoomclick-d3.boot-resid.n-apps\endcsname{1200.0}
\expandafter\def\csname ppt@rev.desc.d6.sspro.p2.T.box.zoomclick-d3.boot-resid.n-apps\endcsname{$+$1200.0}
\expandafter\def\csname val@rev.desc.d6.sspro.p2.T.box.zoomclick-d3.boot-resid.n-boot-valid\endcsname{4000}
\expandafter\def\csname cnt@rev.desc.d6.sspro.p2.T.box.zoomclick-d3.boot-resid.n-boot-valid\endcsname{4{,}000}
\expandafter\def\csname pct@rev.desc.d6.sspro.p2.T.box.zoomclick-d3.boot-resid.n-boot-valid\endcsname{400000.0}
\expandafter\def\csname ppt@rev.desc.d6.sspro.p2.T.box.zoomclick-d3.boot-resid.n-boot-valid\endcsname{$+$400000.0}
\expandafter\def\csname val@rev.desc.d6.sspro.p2.T.box.zoomclick-d3.boot-resid.value\endcsname{0.000}
\expandafter\def\csname cnt@rev.desc.d6.sspro.p2.T.box.zoomclick-d3.boot-resid.value\endcsname{0}
\expandafter\def\csname pct@rev.desc.d6.sspro.p2.T.box.zoomclick-d3.boot-resid.value\endcsname{0.0}
\expandafter\def\csname ppt@rev.desc.d6.sspro.p2.T.box.zoomclick-d3.boot-resid.value\endcsname{0.0}
\expandafter\def\csname val@rev.desc.d6.sspro.p2.T.box.zoomclick-d3.boot-share.ci95.0\endcsname{0.646}
\expandafter\def\csname cnt@rev.desc.d6.sspro.p2.T.box.zoomclick-d3.boot-share.ci95.0\endcsname{1}
\expandafter\def\csname pct@rev.desc.d6.sspro.p2.T.box.zoomclick-d3.boot-share.ci95.0\endcsname{64.6}
\expandafter\def\csname ppt@rev.desc.d6.sspro.p2.T.box.zoomclick-d3.boot-share.ci95.0\endcsname{$+$64.6}
\expandafter\def\csname val@rev.desc.d6.sspro.p2.T.box.zoomclick-d3.boot-share.ci95.1\endcsname{0.812}
\expandafter\def\csname cnt@rev.desc.d6.sspro.p2.T.box.zoomclick-d3.boot-share.ci95.1\endcsname{1}
\expandafter\def\csname pct@rev.desc.d6.sspro.p2.T.box.zoomclick-d3.boot-share.ci95.1\endcsname{81.2}
\expandafter\def\csname ppt@rev.desc.d6.sspro.p2.T.box.zoomclick-d3.boot-share.ci95.1\endcsname{$+$81.2}
\expandafter\def\csname val@rev.desc.d6.sspro.p2.T.box.zoomclick-d3.boot-share.n-apps\endcsname{12}
\expandafter\def\csname cnt@rev.desc.d6.sspro.p2.T.box.zoomclick-d3.boot-share.n-apps\endcsname{12}
\expandafter\def\csname pct@rev.desc.d6.sspro.p2.T.box.zoomclick-d3.boot-share.n-apps\endcsname{1200.0}
\expandafter\def\csname ppt@rev.desc.d6.sspro.p2.T.box.zoomclick-d3.boot-share.n-apps\endcsname{$+$1200.0}
\expandafter\def\csname val@rev.desc.d6.sspro.p2.T.box.zoomclick-d3.boot-share.n-boot-valid\endcsname{4000}
\expandafter\def\csname cnt@rev.desc.d6.sspro.p2.T.box.zoomclick-d3.boot-share.n-boot-valid\endcsname{4{,}000}
\expandafter\def\csname pct@rev.desc.d6.sspro.p2.T.box.zoomclick-d3.boot-share.n-boot-valid\endcsname{400000.0}
\expandafter\def\csname ppt@rev.desc.d6.sspro.p2.T.box.zoomclick-d3.boot-share.n-boot-valid\endcsname{$+$400000.0}
\expandafter\def\csname val@rev.desc.d6.sspro.p2.T.box.zoomclick-d3.boot-share.value\endcsname{0.740}
\expandafter\def\csname cnt@rev.desc.d6.sspro.p2.T.box.zoomclick-d3.boot-share.value\endcsname{1}
\expandafter\def\csname pct@rev.desc.d6.sspro.p2.T.box.zoomclick-d3.boot-share.value\endcsname{74.0}
\expandafter\def\csname ppt@rev.desc.d6.sspro.p2.T.box.zoomclick-d3.boot-share.value\endcsname{$+$74.0}
\expandafter\def\csname val@rev.desc.d6.sspro.p2.T.box.zoomclick-d3.n\endcsname{96}
\expandafter\def\csname cnt@rev.desc.d6.sspro.p2.T.box.zoomclick-d3.n\endcsname{96}
\expandafter\def\csname pct@rev.desc.d6.sspro.p2.T.box.zoomclick-d3.n\endcsname{9600.0}
\expandafter\def\csname ppt@rev.desc.d6.sspro.p2.T.box.zoomclick-d3.n\endcsname{$+$9600.0}
\expandafter\def\csname val@rev.desc.d6.sspro.p2.T.box.zoomclick-d3.resid-agree-stable\endcsname{0.000}
\expandafter\def\csname cnt@rev.desc.d6.sspro.p2.T.box.zoomclick-d3.resid-agree-stable\endcsname{0}
\expandafter\def\csname pct@rev.desc.d6.sspro.p2.T.box.zoomclick-d3.resid-agree-stable\endcsname{0.0}
\expandafter\def\csname ppt@rev.desc.d6.sspro.p2.T.box.zoomclick-d3.resid-agree-stable\endcsname{0.0}
\expandafter\def\csname val@rev.desc.d6.sspro.p2.T.box.zoomclick-d3.resid-disagree\endcsname{0.042}
\expandafter\def\csname cnt@rev.desc.d6.sspro.p2.T.box.zoomclick-d3.resid-disagree\endcsname{0}
\expandafter\def\csname pct@rev.desc.d6.sspro.p2.T.box.zoomclick-d3.resid-disagree\endcsname{4.2}
\expandafter\def\csname ppt@rev.desc.d6.sspro.p2.T.box.zoomclick-d3.resid-disagree\endcsname{$+$4.2}
\expandafter\def\csname val@rev.desc.d6.sspro.p2.T.box.zoomclick-d3.share-agree-stable\endcsname{0.740}
\expandafter\def\csname cnt@rev.desc.d6.sspro.p2.T.box.zoomclick-d3.share-agree-stable\endcsname{1}
\expandafter\def\csname pct@rev.desc.d6.sspro.p2.T.box.zoomclick-d3.share-agree-stable\endcsname{74.0}
\expandafter\def\csname ppt@rev.desc.d6.sspro.p2.T.box.zoomclick-d3.share-agree-stable\endcsname{$+$74.0}
\expandafter\def\csname val@rev.desc.d6.sspro.p2.T.tdiag.catalog-pooled.N-agree-stable\endcsname{330}
\expandafter\def\csname cnt@rev.desc.d6.sspro.p2.T.tdiag.catalog-pooled.N-agree-stable\endcsname{330}
\expandafter\def\csname pct@rev.desc.d6.sspro.p2.T.tdiag.catalog-pooled.N-agree-stable\endcsname{33000.0}
\expandafter\def\csname ppt@rev.desc.d6.sspro.p2.T.tdiag.catalog-pooled.N-agree-stable\endcsname{$+$33000.0}
\expandafter\def\csname val@rev.desc.d6.sspro.p2.T.tdiag.catalog-pooled.N-disagree\endcsname{44}
\expandafter\def\csname cnt@rev.desc.d6.sspro.p2.T.tdiag.catalog-pooled.N-disagree\endcsname{44}
\expandafter\def\csname pct@rev.desc.d6.sspro.p2.T.tdiag.catalog-pooled.N-disagree\endcsname{4400.0}
\expandafter\def\csname ppt@rev.desc.d6.sspro.p2.T.tdiag.catalog-pooled.N-disagree\endcsname{$+$4400.0}
\expandafter\def\csname val@rev.desc.d6.sspro.p2.T.tdiag.catalog-pooled.boot-resid.ci95.0\endcsname{-0.012}
\expandafter\def\csname cnt@rev.desc.d6.sspro.p2.T.tdiag.catalog-pooled.boot-resid.ci95.0\endcsname{0}
\expandafter\def\csname pct@rev.desc.d6.sspro.p2.T.tdiag.catalog-pooled.boot-resid.ci95.0\endcsname{$-$1.2}
\expandafter\def\csname ppt@rev.desc.d6.sspro.p2.T.tdiag.catalog-pooled.boot-resid.ci95.0\endcsname{$-$1.2}
\expandafter\def\csname val@rev.desc.d6.sspro.p2.T.tdiag.catalog-pooled.boot-resid.ci95.1\endcsname{0.029}
\expandafter\def\csname cnt@rev.desc.d6.sspro.p2.T.tdiag.catalog-pooled.boot-resid.ci95.1\endcsname{0}
\expandafter\def\csname pct@rev.desc.d6.sspro.p2.T.tdiag.catalog-pooled.boot-resid.ci95.1\endcsname{2.9}
\expandafter\def\csname ppt@rev.desc.d6.sspro.p2.T.tdiag.catalog-pooled.boot-resid.ci95.1\endcsname{$+$2.9}
\expandafter\def\csname val@rev.desc.d6.sspro.p2.T.tdiag.catalog-pooled.boot-resid.n-apps\endcsname{12}
\expandafter\def\csname cnt@rev.desc.d6.sspro.p2.T.tdiag.catalog-pooled.boot-resid.n-apps\endcsname{12}
\expandafter\def\csname pct@rev.desc.d6.sspro.p2.T.tdiag.catalog-pooled.boot-resid.n-apps\endcsname{1200.0}
\expandafter\def\csname ppt@rev.desc.d6.sspro.p2.T.tdiag.catalog-pooled.boot-resid.n-apps\endcsname{$+$1200.0}
\expandafter\def\csname val@rev.desc.d6.sspro.p2.T.tdiag.catalog-pooled.boot-resid.n-boot-valid\endcsname{4000}
\expandafter\def\csname cnt@rev.desc.d6.sspro.p2.T.tdiag.catalog-pooled.boot-resid.n-boot-valid\endcsname{4{,}000}
\expandafter\def\csname pct@rev.desc.d6.sspro.p2.T.tdiag.catalog-pooled.boot-resid.n-boot-valid\endcsname{400000.0}
\expandafter\def\csname ppt@rev.desc.d6.sspro.p2.T.tdiag.catalog-pooled.boot-resid.n-boot-valid\endcsname{$+$400000.0}
\expandafter\def\csname val@rev.desc.d6.sspro.p2.T.tdiag.catalog-pooled.boot-resid.value\endcsname{0.009}
\expandafter\def\csname cnt@rev.desc.d6.sspro.p2.T.tdiag.catalog-pooled.boot-resid.value\endcsname{0}
\expandafter\def\csname pct@rev.desc.d6.sspro.p2.T.tdiag.catalog-pooled.boot-resid.value\endcsname{0.9}
\expandafter\def\csname ppt@rev.desc.d6.sspro.p2.T.tdiag.catalog-pooled.boot-resid.value\endcsname{$+$0.9}
\expandafter\def\csname val@rev.desc.d6.sspro.p2.T.tdiag.catalog-pooled.boot-share.ci95.0\endcsname{0.799}
\expandafter\def\csname cnt@rev.desc.d6.sspro.p2.T.tdiag.catalog-pooled.boot-share.ci95.0\endcsname{1}
\expandafter\def\csname pct@rev.desc.d6.sspro.p2.T.tdiag.catalog-pooled.boot-share.ci95.0\endcsname{79.9}
\expandafter\def\csname ppt@rev.desc.d6.sspro.p2.T.tdiag.catalog-pooled.boot-share.ci95.0\endcsname{$+$79.9}
\expandafter\def\csname val@rev.desc.d6.sspro.p2.T.tdiag.catalog-pooled.boot-share.ci95.1\endcsname{0.911}
\expandafter\def\csname cnt@rev.desc.d6.sspro.p2.T.tdiag.catalog-pooled.boot-share.ci95.1\endcsname{1}
\expandafter\def\csname pct@rev.desc.d6.sspro.p2.T.tdiag.catalog-pooled.boot-share.ci95.1\endcsname{91.1}
\expandafter\def\csname ppt@rev.desc.d6.sspro.p2.T.tdiag.catalog-pooled.boot-share.ci95.1\endcsname{$+$91.1}
\expandafter\def\csname val@rev.desc.d6.sspro.p2.T.tdiag.catalog-pooled.boot-share.n-apps\endcsname{12}
\expandafter\def\csname cnt@rev.desc.d6.sspro.p2.T.tdiag.catalog-pooled.boot-share.n-apps\endcsname{12}
\expandafter\def\csname pct@rev.desc.d6.sspro.p2.T.tdiag.catalog-pooled.boot-share.n-apps\endcsname{1200.0}
\expandafter\def\csname ppt@rev.desc.d6.sspro.p2.T.tdiag.catalog-pooled.boot-share.n-apps\endcsname{$+$1200.0}
\expandafter\def\csname val@rev.desc.d6.sspro.p2.T.tdiag.catalog-pooled.boot-share.n-boot-valid\endcsname{4000}
\expandafter\def\csname cnt@rev.desc.d6.sspro.p2.T.tdiag.catalog-pooled.boot-share.n-boot-valid\endcsname{4{,}000}
\expandafter\def\csname pct@rev.desc.d6.sspro.p2.T.tdiag.catalog-pooled.boot-share.n-boot-valid\endcsname{400000.0}
\expandafter\def\csname ppt@rev.desc.d6.sspro.p2.T.tdiag.catalog-pooled.boot-share.n-boot-valid\endcsname{$+$400000.0}
\expandafter\def\csname val@rev.desc.d6.sspro.p2.T.tdiag.catalog-pooled.boot-share.value\endcsname{0.859}
\expandafter\def\csname cnt@rev.desc.d6.sspro.p2.T.tdiag.catalog-pooled.boot-share.value\endcsname{1}
\expandafter\def\csname pct@rev.desc.d6.sspro.p2.T.tdiag.catalog-pooled.boot-share.value\endcsname{85.9}
\expandafter\def\csname ppt@rev.desc.d6.sspro.p2.T.tdiag.catalog-pooled.boot-share.value\endcsname{$+$85.9}
\expandafter\def\csname val@rev.desc.d6.sspro.p2.T.tdiag.catalog-pooled.gap\endcsname{0.082}
\expandafter\def\csname cnt@rev.desc.d6.sspro.p2.T.tdiag.catalog-pooled.gap\endcsname{0}
\expandafter\def\csname pct@rev.desc.d6.sspro.p2.T.tdiag.catalog-pooled.gap\endcsname{8.2}
\expandafter\def\csname ppt@rev.desc.d6.sspro.p2.T.tdiag.catalog-pooled.gap\endcsname{$+$8.2}
\expandafter\def\csname val@rev.desc.d6.sspro.p2.T.tdiag.catalog-pooled.n\endcsname{384}
\expandafter\def\csname cnt@rev.desc.d6.sspro.p2.T.tdiag.catalog-pooled.n\endcsname{384}
\expandafter\def\csname pct@rev.desc.d6.sspro.p2.T.tdiag.catalog-pooled.n\endcsname{38400.0}
\expandafter\def\csname ppt@rev.desc.d6.sspro.p2.T.tdiag.catalog-pooled.n\endcsname{$+$38400.0}
\expandafter\def\csname val@rev.desc.d6.sspro.p2.T.tdiag.catalog-pooled.resid-agree-stable\endcsname{0.009}
\expandafter\def\csname cnt@rev.desc.d6.sspro.p2.T.tdiag.catalog-pooled.resid-agree-stable\endcsname{0}
\expandafter\def\csname pct@rev.desc.d6.sspro.p2.T.tdiag.catalog-pooled.resid-agree-stable\endcsname{0.9}
\expandafter\def\csname ppt@rev.desc.d6.sspro.p2.T.tdiag.catalog-pooled.resid-agree-stable\endcsname{$+$0.9}
\expandafter\def\csname val@rev.desc.d6.sspro.p2.T.tdiag.catalog-pooled.resid-disagree\endcsname{0.091}
\expandafter\def\csname cnt@rev.desc.d6.sspro.p2.T.tdiag.catalog-pooled.resid-disagree\endcsname{0}
\expandafter\def\csname pct@rev.desc.d6.sspro.p2.T.tdiag.catalog-pooled.resid-disagree\endcsname{9.1}
\expandafter\def\csname ppt@rev.desc.d6.sspro.p2.T.tdiag.catalog-pooled.resid-disagree\endcsname{$+$9.1}
\expandafter\def\csname val@rev.desc.d6.sspro.p2.T.tdiag.catalog-pooled.share-agree-stable\endcsname{0.859}
\expandafter\def\csname cnt@rev.desc.d6.sspro.p2.T.tdiag.catalog-pooled.share-agree-stable\endcsname{1}
\expandafter\def\csname pct@rev.desc.d6.sspro.p2.T.tdiag.catalog-pooled.share-agree-stable\endcsname{85.9}
\expandafter\def\csname ppt@rev.desc.d6.sspro.p2.T.tdiag.catalog-pooled.share-agree-stable\endcsname{$+$85.9}
\expandafter\def\csname val@rev.desc.d6.sspro.p2.T.tdiag.center-d1.N-agree-stable\endcsname{86}
\expandafter\def\csname cnt@rev.desc.d6.sspro.p2.T.tdiag.center-d1.N-agree-stable\endcsname{86}
\expandafter\def\csname pct@rev.desc.d6.sspro.p2.T.tdiag.center-d1.N-agree-stable\endcsname{8600.0}
\expandafter\def\csname ppt@rev.desc.d6.sspro.p2.T.tdiag.center-d1.N-agree-stable\endcsname{$+$8600.0}
\expandafter\def\csname val@rev.desc.d6.sspro.p2.T.tdiag.center-d1.N-disagree\endcsname{9}
\expandafter\def\csname cnt@rev.desc.d6.sspro.p2.T.tdiag.center-d1.N-disagree\endcsname{9}
\expandafter\def\csname pct@rev.desc.d6.sspro.p2.T.tdiag.center-d1.N-disagree\endcsname{900.0}
\expandafter\def\csname ppt@rev.desc.d6.sspro.p2.T.tdiag.center-d1.N-disagree\endcsname{$+$900.0}
\expandafter\def\csname val@rev.desc.d6.sspro.p2.T.tdiag.center-d1.boot-resid.ci95.0\endcsname{-0.024}
\expandafter\def\csname cnt@rev.desc.d6.sspro.p2.T.tdiag.center-d1.boot-resid.ci95.0\endcsname{0}
\expandafter\def\csname pct@rev.desc.d6.sspro.p2.T.tdiag.center-d1.boot-resid.ci95.0\endcsname{$-$2.4}
\expandafter\def\csname ppt@rev.desc.d6.sspro.p2.T.tdiag.center-d1.boot-resid.ci95.0\endcsname{$-$2.4}
\expandafter\def\csname val@rev.desc.d6.sspro.p2.T.tdiag.center-d1.boot-resid.ci95.1\endcsname{0.049}
\expandafter\def\csname cnt@rev.desc.d6.sspro.p2.T.tdiag.center-d1.boot-resid.ci95.1\endcsname{0}
\expandafter\def\csname pct@rev.desc.d6.sspro.p2.T.tdiag.center-d1.boot-resid.ci95.1\endcsname{4.9}
\expandafter\def\csname ppt@rev.desc.d6.sspro.p2.T.tdiag.center-d1.boot-resid.ci95.1\endcsname{$+$4.9}
\expandafter\def\csname val@rev.desc.d6.sspro.p2.T.tdiag.center-d1.boot-resid.n-apps\endcsname{12}
\expandafter\def\csname cnt@rev.desc.d6.sspro.p2.T.tdiag.center-d1.boot-resid.n-apps\endcsname{12}
\expandafter\def\csname pct@rev.desc.d6.sspro.p2.T.tdiag.center-d1.boot-resid.n-apps\endcsname{1200.0}
\expandafter\def\csname ppt@rev.desc.d6.sspro.p2.T.tdiag.center-d1.boot-resid.n-apps\endcsname{$+$1200.0}
\expandafter\def\csname val@rev.desc.d6.sspro.p2.T.tdiag.center-d1.boot-resid.n-boot-valid\endcsname{4000}
\expandafter\def\csname cnt@rev.desc.d6.sspro.p2.T.tdiag.center-d1.boot-resid.n-boot-valid\endcsname{4{,}000}
\expandafter\def\csname pct@rev.desc.d6.sspro.p2.T.tdiag.center-d1.boot-resid.n-boot-valid\endcsname{400000.0}
\expandafter\def\csname ppt@rev.desc.d6.sspro.p2.T.tdiag.center-d1.boot-resid.n-boot-valid\endcsname{$+$400000.0}
\expandafter\def\csname val@rev.desc.d6.sspro.p2.T.tdiag.center-d1.boot-resid.value\endcsname{0.012}
\expandafter\def\csname cnt@rev.desc.d6.sspro.p2.T.tdiag.center-d1.boot-resid.value\endcsname{0}
\expandafter\def\csname pct@rev.desc.d6.sspro.p2.T.tdiag.center-d1.boot-resid.value\endcsname{1.2}
\expandafter\def\csname ppt@rev.desc.d6.sspro.p2.T.tdiag.center-d1.boot-resid.value\endcsname{$+$1.2}
\expandafter\def\csname val@rev.desc.d6.sspro.p2.T.tdiag.center-d1.boot-share.ci95.0\endcsname{0.833}
\expandafter\def\csname cnt@rev.desc.d6.sspro.p2.T.tdiag.center-d1.boot-share.ci95.0\endcsname{1}
\expandafter\def\csname pct@rev.desc.d6.sspro.p2.T.tdiag.center-d1.boot-share.ci95.0\endcsname{83.3}
\expandafter\def\csname ppt@rev.desc.d6.sspro.p2.T.tdiag.center-d1.boot-share.ci95.0\endcsname{$+$83.3}
\expandafter\def\csname val@rev.desc.d6.sspro.p2.T.tdiag.center-d1.boot-share.ci95.1\endcsname{0.948}
\expandafter\def\csname cnt@rev.desc.d6.sspro.p2.T.tdiag.center-d1.boot-share.ci95.1\endcsname{1}
\expandafter\def\csname pct@rev.desc.d6.sspro.p2.T.tdiag.center-d1.boot-share.ci95.1\endcsname{94.8}
\expandafter\def\csname ppt@rev.desc.d6.sspro.p2.T.tdiag.center-d1.boot-share.ci95.1\endcsname{$+$94.8}
\expandafter\def\csname val@rev.desc.d6.sspro.p2.T.tdiag.center-d1.boot-share.n-apps\endcsname{12}
\expandafter\def\csname cnt@rev.desc.d6.sspro.p2.T.tdiag.center-d1.boot-share.n-apps\endcsname{12}
\expandafter\def\csname pct@rev.desc.d6.sspro.p2.T.tdiag.center-d1.boot-share.n-apps\endcsname{1200.0}
\expandafter\def\csname ppt@rev.desc.d6.sspro.p2.T.tdiag.center-d1.boot-share.n-apps\endcsname{$+$1200.0}
\expandafter\def\csname val@rev.desc.d6.sspro.p2.T.tdiag.center-d1.boot-share.n-boot-valid\endcsname{4000}
\expandafter\def\csname cnt@rev.desc.d6.sspro.p2.T.tdiag.center-d1.boot-share.n-boot-valid\endcsname{4{,}000}
\expandafter\def\csname pct@rev.desc.d6.sspro.p2.T.tdiag.center-d1.boot-share.n-boot-valid\endcsname{400000.0}
\expandafter\def\csname ppt@rev.desc.d6.sspro.p2.T.tdiag.center-d1.boot-share.n-boot-valid\endcsname{$+$400000.0}
\expandafter\def\csname val@rev.desc.d6.sspro.p2.T.tdiag.center-d1.boot-share.value\endcsname{0.896}
\expandafter\def\csname cnt@rev.desc.d6.sspro.p2.T.tdiag.center-d1.boot-share.value\endcsname{1}
\expandafter\def\csname pct@rev.desc.d6.sspro.p2.T.tdiag.center-d1.boot-share.value\endcsname{89.6}
\expandafter\def\csname ppt@rev.desc.d6.sspro.p2.T.tdiag.center-d1.boot-share.value\endcsname{$+$89.6}
\expandafter\def\csname val@rev.desc.d6.sspro.p2.T.tdiag.center-d1.n\endcsname{96}
\expandafter\def\csname cnt@rev.desc.d6.sspro.p2.T.tdiag.center-d1.n\endcsname{96}
\expandafter\def\csname pct@rev.desc.d6.sspro.p2.T.tdiag.center-d1.n\endcsname{9600.0}
\expandafter\def\csname ppt@rev.desc.d6.sspro.p2.T.tdiag.center-d1.n\endcsname{$+$9600.0}
\expandafter\def\csname val@rev.desc.d6.sspro.p2.T.tdiag.center-d1.resid-agree-stable\endcsname{0.012}
\expandafter\def\csname cnt@rev.desc.d6.sspro.p2.T.tdiag.center-d1.resid-agree-stable\endcsname{0}
\expandafter\def\csname pct@rev.desc.d6.sspro.p2.T.tdiag.center-d1.resid-agree-stable\endcsname{1.2}
\expandafter\def\csname ppt@rev.desc.d6.sspro.p2.T.tdiag.center-d1.resid-agree-stable\endcsname{$+$1.2}
\expandafter\def\csname val@rev.desc.d6.sspro.p2.T.tdiag.center-d1.resid-disagree\endcsname{0.222}
\expandafter\def\csname cnt@rev.desc.d6.sspro.p2.T.tdiag.center-d1.resid-disagree\endcsname{0}
\expandafter\def\csname pct@rev.desc.d6.sspro.p2.T.tdiag.center-d1.resid-disagree\endcsname{22.2}
\expandafter\def\csname ppt@rev.desc.d6.sspro.p2.T.tdiag.center-d1.resid-disagree\endcsname{$+$22.2}
\expandafter\def\csname val@rev.desc.d6.sspro.p2.T.tdiag.center-d1.share-agree-stable\endcsname{0.896}
\expandafter\def\csname cnt@rev.desc.d6.sspro.p2.T.tdiag.center-d1.share-agree-stable\endcsname{1}
\expandafter\def\csname pct@rev.desc.d6.sspro.p2.T.tdiag.center-d1.share-agree-stable\endcsname{89.6}
\expandafter\def\csname ppt@rev.desc.d6.sspro.p2.T.tdiag.center-d1.share-agree-stable\endcsname{$+$89.6}
\expandafter\def\csname val@rev.desc.d6.sspro.p2.T.tdiag.context-d1.N-agree-stable\endcsname{84}
\expandafter\def\csname cnt@rev.desc.d6.sspro.p2.T.tdiag.context-d1.N-agree-stable\endcsname{84}
\expandafter\def\csname pct@rev.desc.d6.sspro.p2.T.tdiag.context-d1.N-agree-stable\endcsname{8400.0}
\expandafter\def\csname ppt@rev.desc.d6.sspro.p2.T.tdiag.context-d1.N-agree-stable\endcsname{$+$8400.0}
\expandafter\def\csname val@rev.desc.d6.sspro.p2.T.tdiag.context-d1.N-disagree\endcsname{10}
\expandafter\def\csname cnt@rev.desc.d6.sspro.p2.T.tdiag.context-d1.N-disagree\endcsname{10}
\expandafter\def\csname pct@rev.desc.d6.sspro.p2.T.tdiag.context-d1.N-disagree\endcsname{1000.0}
\expandafter\def\csname ppt@rev.desc.d6.sspro.p2.T.tdiag.context-d1.N-disagree\endcsname{$+$1000.0}
\expandafter\def\csname val@rev.desc.d6.sspro.p2.T.tdiag.context-d1.boot-resid.ci95.0\endcsname{-0.070}
\expandafter\def\csname cnt@rev.desc.d6.sspro.p2.T.tdiag.context-d1.boot-resid.ci95.0\endcsname{0}
\expandafter\def\csname pct@rev.desc.d6.sspro.p2.T.tdiag.context-d1.boot-resid.ci95.0\endcsname{$-$7.0}
\expandafter\def\csname ppt@rev.desc.d6.sspro.p2.T.tdiag.context-d1.boot-resid.ci95.0\endcsname{$-$7.0}
\expandafter\def\csname val@rev.desc.d6.sspro.p2.T.tdiag.context-d1.boot-resid.ci95.1\endcsname{0.023}
\expandafter\def\csname cnt@rev.desc.d6.sspro.p2.T.tdiag.context-d1.boot-resid.ci95.1\endcsname{0}
\expandafter\def\csname pct@rev.desc.d6.sspro.p2.T.tdiag.context-d1.boot-resid.ci95.1\endcsname{2.3}
\expandafter\def\csname ppt@rev.desc.d6.sspro.p2.T.tdiag.context-d1.boot-resid.ci95.1\endcsname{$+$2.3}
\expandafter\def\csname val@rev.desc.d6.sspro.p2.T.tdiag.context-d1.boot-resid.n-apps\endcsname{12}
\expandafter\def\csname cnt@rev.desc.d6.sspro.p2.T.tdiag.context-d1.boot-resid.n-apps\endcsname{12}
\expandafter\def\csname pct@rev.desc.d6.sspro.p2.T.tdiag.context-d1.boot-resid.n-apps\endcsname{1200.0}
\expandafter\def\csname ppt@rev.desc.d6.sspro.p2.T.tdiag.context-d1.boot-resid.n-apps\endcsname{$+$1200.0}
\expandafter\def\csname val@rev.desc.d6.sspro.p2.T.tdiag.context-d1.boot-resid.n-boot-valid\endcsname{4000}
\expandafter\def\csname cnt@rev.desc.d6.sspro.p2.T.tdiag.context-d1.boot-resid.n-boot-valid\endcsname{4{,}000}
\expandafter\def\csname pct@rev.desc.d6.sspro.p2.T.tdiag.context-d1.boot-resid.n-boot-valid\endcsname{400000.0}
\expandafter\def\csname ppt@rev.desc.d6.sspro.p2.T.tdiag.context-d1.boot-resid.n-boot-valid\endcsname{$+$400000.0}
\expandafter\def\csname val@rev.desc.d6.sspro.p2.T.tdiag.context-d1.boot-resid.value\endcsname{-0.024}
\expandafter\def\csname cnt@rev.desc.d6.sspro.p2.T.tdiag.context-d1.boot-resid.value\endcsname{0}
\expandafter\def\csname pct@rev.desc.d6.sspro.p2.T.tdiag.context-d1.boot-resid.value\endcsname{$-$2.4}
\expandafter\def\csname ppt@rev.desc.d6.sspro.p2.T.tdiag.context-d1.boot-resid.value\endcsname{$-$2.4}
\expandafter\def\csname val@rev.desc.d6.sspro.p2.T.tdiag.context-d1.boot-share.ci95.0\endcsname{0.812}
\expandafter\def\csname cnt@rev.desc.d6.sspro.p2.T.tdiag.context-d1.boot-share.ci95.0\endcsname{1}
\expandafter\def\csname pct@rev.desc.d6.sspro.p2.T.tdiag.context-d1.boot-share.ci95.0\endcsname{81.2}
\expandafter\def\csname ppt@rev.desc.d6.sspro.p2.T.tdiag.context-d1.boot-share.ci95.0\endcsname{$+$81.2}
\expandafter\def\csname val@rev.desc.d6.sspro.p2.T.tdiag.context-d1.boot-share.ci95.1\endcsname{0.927}
\expandafter\def\csname cnt@rev.desc.d6.sspro.p2.T.tdiag.context-d1.boot-share.ci95.1\endcsname{1}
\expandafter\def\csname pct@rev.desc.d6.sspro.p2.T.tdiag.context-d1.boot-share.ci95.1\endcsname{92.7}
\expandafter\def\csname ppt@rev.desc.d6.sspro.p2.T.tdiag.context-d1.boot-share.ci95.1\endcsname{$+$92.7}
\expandafter\def\csname val@rev.desc.d6.sspro.p2.T.tdiag.context-d1.boot-share.n-apps\endcsname{12}
\expandafter\def\csname cnt@rev.desc.d6.sspro.p2.T.tdiag.context-d1.boot-share.n-apps\endcsname{12}
\expandafter\def\csname pct@rev.desc.d6.sspro.p2.T.tdiag.context-d1.boot-share.n-apps\endcsname{1200.0}
\expandafter\def\csname ppt@rev.desc.d6.sspro.p2.T.tdiag.context-d1.boot-share.n-apps\endcsname{$+$1200.0}
\expandafter\def\csname val@rev.desc.d6.sspro.p2.T.tdiag.context-d1.boot-share.n-boot-valid\endcsname{4000}
\expandafter\def\csname cnt@rev.desc.d6.sspro.p2.T.tdiag.context-d1.boot-share.n-boot-valid\endcsname{4{,}000}
\expandafter\def\csname pct@rev.desc.d6.sspro.p2.T.tdiag.context-d1.boot-share.n-boot-valid\endcsname{400000.0}
\expandafter\def\csname ppt@rev.desc.d6.sspro.p2.T.tdiag.context-d1.boot-share.n-boot-valid\endcsname{$+$400000.0}
\expandafter\def\csname val@rev.desc.d6.sspro.p2.T.tdiag.context-d1.boot-share.value\endcsname{0.875}
\expandafter\def\csname cnt@rev.desc.d6.sspro.p2.T.tdiag.context-d1.boot-share.value\endcsname{1}
\expandafter\def\csname pct@rev.desc.d6.sspro.p2.T.tdiag.context-d1.boot-share.value\endcsname{87.5}
\expandafter\def\csname ppt@rev.desc.d6.sspro.p2.T.tdiag.context-d1.boot-share.value\endcsname{$+$87.5}
\expandafter\def\csname val@rev.desc.d6.sspro.p2.T.tdiag.context-d1.n\endcsname{96}
\expandafter\def\csname cnt@rev.desc.d6.sspro.p2.T.tdiag.context-d1.n\endcsname{96}
\expandafter\def\csname pct@rev.desc.d6.sspro.p2.T.tdiag.context-d1.n\endcsname{9600.0}
\expandafter\def\csname ppt@rev.desc.d6.sspro.p2.T.tdiag.context-d1.n\endcsname{$+$9600.0}
\expandafter\def\csname val@rev.desc.d6.sspro.p2.T.tdiag.context-d1.resid-agree-stable\endcsname{-0.024}
\expandafter\def\csname cnt@rev.desc.d6.sspro.p2.T.tdiag.context-d1.resid-agree-stable\endcsname{0}
\expandafter\def\csname pct@rev.desc.d6.sspro.p2.T.tdiag.context-d1.resid-agree-stable\endcsname{$-$2.4}
\expandafter\def\csname ppt@rev.desc.d6.sspro.p2.T.tdiag.context-d1.resid-agree-stable\endcsname{$-$2.4}
\expandafter\def\csname val@rev.desc.d6.sspro.p2.T.tdiag.context-d1.resid-disagree\endcsname{0.000}
\expandafter\def\csname cnt@rev.desc.d6.sspro.p2.T.tdiag.context-d1.resid-disagree\endcsname{0}
\expandafter\def\csname pct@rev.desc.d6.sspro.p2.T.tdiag.context-d1.resid-disagree\endcsname{0.0}
\expandafter\def\csname ppt@rev.desc.d6.sspro.p2.T.tdiag.context-d1.resid-disagree\endcsname{0.0}
\expandafter\def\csname val@rev.desc.d6.sspro.p2.T.tdiag.context-d1.share-agree-stable\endcsname{0.875}
\expandafter\def\csname cnt@rev.desc.d6.sspro.p2.T.tdiag.context-d1.share-agree-stable\endcsname{1}
\expandafter\def\csname pct@rev.desc.d6.sspro.p2.T.tdiag.context-d1.share-agree-stable\endcsname{87.5}
\expandafter\def\csname ppt@rev.desc.d6.sspro.p2.T.tdiag.context-d1.share-agree-stable\endcsname{$+$87.5}
\expandafter\def\csname val@rev.desc.d6.sspro.p2.T.tdiag.dual-d2.N-agree-stable\endcsname{81}
\expandafter\def\csname cnt@rev.desc.d6.sspro.p2.T.tdiag.dual-d2.N-agree-stable\endcsname{81}
\expandafter\def\csname pct@rev.desc.d6.sspro.p2.T.tdiag.dual-d2.N-agree-stable\endcsname{8100.0}
\expandafter\def\csname ppt@rev.desc.d6.sspro.p2.T.tdiag.dual-d2.N-agree-stable\endcsname{$+$8100.0}
\expandafter\def\csname val@rev.desc.d6.sspro.p2.T.tdiag.dual-d2.N-disagree\endcsname{13}
\expandafter\def\csname cnt@rev.desc.d6.sspro.p2.T.tdiag.dual-d2.N-disagree\endcsname{13}
\expandafter\def\csname pct@rev.desc.d6.sspro.p2.T.tdiag.dual-d2.N-disagree\endcsname{1300.0}
\expandafter\def\csname ppt@rev.desc.d6.sspro.p2.T.tdiag.dual-d2.N-disagree\endcsname{$+$1300.0}
\expandafter\def\csname val@rev.desc.d6.sspro.p2.T.tdiag.dual-d2.boot-resid.ci95.0\endcsname{0.000}
\expandafter\def\csname cnt@rev.desc.d6.sspro.p2.T.tdiag.dual-d2.boot-resid.ci95.0\endcsname{0}
\expandafter\def\csname pct@rev.desc.d6.sspro.p2.T.tdiag.dual-d2.boot-resid.ci95.0\endcsname{0.0}
\expandafter\def\csname ppt@rev.desc.d6.sspro.p2.T.tdiag.dual-d2.boot-resid.ci95.0\endcsname{0.0}
\expandafter\def\csname val@rev.desc.d6.sspro.p2.T.tdiag.dual-d2.boot-resid.ci95.1\endcsname{0.073}
\expandafter\def\csname cnt@rev.desc.d6.sspro.p2.T.tdiag.dual-d2.boot-resid.ci95.1\endcsname{0}
\expandafter\def\csname pct@rev.desc.d6.sspro.p2.T.tdiag.dual-d2.boot-resid.ci95.1\endcsname{7.3}
\expandafter\def\csname ppt@rev.desc.d6.sspro.p2.T.tdiag.dual-d2.boot-resid.ci95.1\endcsname{$+$7.3}
\expandafter\def\csname val@rev.desc.d6.sspro.p2.T.tdiag.dual-d2.boot-resid.n-apps\endcsname{12}
\expandafter\def\csname cnt@rev.desc.d6.sspro.p2.T.tdiag.dual-d2.boot-resid.n-apps\endcsname{12}
\expandafter\def\csname pct@rev.desc.d6.sspro.p2.T.tdiag.dual-d2.boot-resid.n-apps\endcsname{1200.0}
\expandafter\def\csname ppt@rev.desc.d6.sspro.p2.T.tdiag.dual-d2.boot-resid.n-apps\endcsname{$+$1200.0}
\expandafter\def\csname val@rev.desc.d6.sspro.p2.T.tdiag.dual-d2.boot-resid.n-boot-valid\endcsname{4000}
\expandafter\def\csname cnt@rev.desc.d6.sspro.p2.T.tdiag.dual-d2.boot-resid.n-boot-valid\endcsname{4{,}000}
\expandafter\def\csname pct@rev.desc.d6.sspro.p2.T.tdiag.dual-d2.boot-resid.n-boot-valid\endcsname{400000.0}
\expandafter\def\csname ppt@rev.desc.d6.sspro.p2.T.tdiag.dual-d2.boot-resid.n-boot-valid\endcsname{$+$400000.0}
\expandafter\def\csname val@rev.desc.d6.sspro.p2.T.tdiag.dual-d2.boot-resid.value\endcsname{0.037}
\expandafter\def\csname cnt@rev.desc.d6.sspro.p2.T.tdiag.dual-d2.boot-resid.value\endcsname{0}
\expandafter\def\csname pct@rev.desc.d6.sspro.p2.T.tdiag.dual-d2.boot-resid.value\endcsname{3.7}
\expandafter\def\csname ppt@rev.desc.d6.sspro.p2.T.tdiag.dual-d2.boot-resid.value\endcsname{$+$3.7}
\expandafter\def\csname val@rev.desc.d6.sspro.p2.T.tdiag.dual-d2.boot-share.ci95.0\endcsname{0.781}
\expandafter\def\csname cnt@rev.desc.d6.sspro.p2.T.tdiag.dual-d2.boot-share.ci95.0\endcsname{1}
\expandafter\def\csname pct@rev.desc.d6.sspro.p2.T.tdiag.dual-d2.boot-share.ci95.0\endcsname{78.1}
\expandafter\def\csname ppt@rev.desc.d6.sspro.p2.T.tdiag.dual-d2.boot-share.ci95.0\endcsname{$+$78.1}
\expandafter\def\csname val@rev.desc.d6.sspro.p2.T.tdiag.dual-d2.boot-share.ci95.1\endcsname{0.896}
\expandafter\def\csname cnt@rev.desc.d6.sspro.p2.T.tdiag.dual-d2.boot-share.ci95.1\endcsname{1}
\expandafter\def\csname pct@rev.desc.d6.sspro.p2.T.tdiag.dual-d2.boot-share.ci95.1\endcsname{89.6}
\expandafter\def\csname ppt@rev.desc.d6.sspro.p2.T.tdiag.dual-d2.boot-share.ci95.1\endcsname{$+$89.6}
\expandafter\def\csname val@rev.desc.d6.sspro.p2.T.tdiag.dual-d2.boot-share.n-apps\endcsname{12}
\expandafter\def\csname cnt@rev.desc.d6.sspro.p2.T.tdiag.dual-d2.boot-share.n-apps\endcsname{12}
\expandafter\def\csname pct@rev.desc.d6.sspro.p2.T.tdiag.dual-d2.boot-share.n-apps\endcsname{1200.0}
\expandafter\def\csname ppt@rev.desc.d6.sspro.p2.T.tdiag.dual-d2.boot-share.n-apps\endcsname{$+$1200.0}
\expandafter\def\csname val@rev.desc.d6.sspro.p2.T.tdiag.dual-d2.boot-share.n-boot-valid\endcsname{4000}
\expandafter\def\csname cnt@rev.desc.d6.sspro.p2.T.tdiag.dual-d2.boot-share.n-boot-valid\endcsname{4{,}000}
\expandafter\def\csname pct@rev.desc.d6.sspro.p2.T.tdiag.dual-d2.boot-share.n-boot-valid\endcsname{400000.0}
\expandafter\def\csname ppt@rev.desc.d6.sspro.p2.T.tdiag.dual-d2.boot-share.n-boot-valid\endcsname{$+$400000.0}
\expandafter\def\csname val@rev.desc.d6.sspro.p2.T.tdiag.dual-d2.boot-share.value\endcsname{0.844}
\expandafter\def\csname cnt@rev.desc.d6.sspro.p2.T.tdiag.dual-d2.boot-share.value\endcsname{1}
\expandafter\def\csname pct@rev.desc.d6.sspro.p2.T.tdiag.dual-d2.boot-share.value\endcsname{84.4}
\expandafter\def\csname ppt@rev.desc.d6.sspro.p2.T.tdiag.dual-d2.boot-share.value\endcsname{$+$84.4}
\expandafter\def\csname val@rev.desc.d6.sspro.p2.T.tdiag.dual-d2.n\endcsname{96}
\expandafter\def\csname cnt@rev.desc.d6.sspro.p2.T.tdiag.dual-d2.n\endcsname{96}
\expandafter\def\csname pct@rev.desc.d6.sspro.p2.T.tdiag.dual-d2.n\endcsname{9600.0}
\expandafter\def\csname ppt@rev.desc.d6.sspro.p2.T.tdiag.dual-d2.n\endcsname{$+$9600.0}
\expandafter\def\csname val@rev.desc.d6.sspro.p2.T.tdiag.dual-d2.resid-agree-stable\endcsname{0.037}
\expandafter\def\csname cnt@rev.desc.d6.sspro.p2.T.tdiag.dual-d2.resid-agree-stable\endcsname{0}
\expandafter\def\csname pct@rev.desc.d6.sspro.p2.T.tdiag.dual-d2.resid-agree-stable\endcsname{3.7}
\expandafter\def\csname ppt@rev.desc.d6.sspro.p2.T.tdiag.dual-d2.resid-agree-stable\endcsname{$+$3.7}
\expandafter\def\csname val@rev.desc.d6.sspro.p2.T.tdiag.dual-d2.resid-disagree\endcsname{0.154}
\expandafter\def\csname cnt@rev.desc.d6.sspro.p2.T.tdiag.dual-d2.resid-disagree\endcsname{0}
\expandafter\def\csname pct@rev.desc.d6.sspro.p2.T.tdiag.dual-d2.resid-disagree\endcsname{15.4}
\expandafter\def\csname ppt@rev.desc.d6.sspro.p2.T.tdiag.dual-d2.resid-disagree\endcsname{$+$15.4}
\expandafter\def\csname val@rev.desc.d6.sspro.p2.T.tdiag.dual-d2.share-agree-stable\endcsname{0.844}
\expandafter\def\csname cnt@rev.desc.d6.sspro.p2.T.tdiag.dual-d2.share-agree-stable\endcsname{1}
\expandafter\def\csname pct@rev.desc.d6.sspro.p2.T.tdiag.dual-d2.share-agree-stable\endcsname{84.4}
\expandafter\def\csname ppt@rev.desc.d6.sspro.p2.T.tdiag.dual-d2.share-agree-stable\endcsname{$+$84.4}
\expandafter\def\csname val@rev.desc.d6.sspro.p2.T.tdiag.grid2-d3.N-agree-stable\endcsname{79}
\expandafter\def\csname cnt@rev.desc.d6.sspro.p2.T.tdiag.grid2-d3.N-agree-stable\endcsname{79}
\expandafter\def\csname pct@rev.desc.d6.sspro.p2.T.tdiag.grid2-d3.N-agree-stable\endcsname{7900.0}
\expandafter\def\csname ppt@rev.desc.d6.sspro.p2.T.tdiag.grid2-d3.N-agree-stable\endcsname{$+$7900.0}
\expandafter\def\csname val@rev.desc.d6.sspro.p2.T.tdiag.grid2-d3.N-disagree\endcsname{12}
\expandafter\def\csname cnt@rev.desc.d6.sspro.p2.T.tdiag.grid2-d3.N-disagree\endcsname{12}
\expandafter\def\csname pct@rev.desc.d6.sspro.p2.T.tdiag.grid2-d3.N-disagree\endcsname{1200.0}
\expandafter\def\csname ppt@rev.desc.d6.sspro.p2.T.tdiag.grid2-d3.N-disagree\endcsname{$+$1200.0}
\expandafter\def\csname val@rev.desc.d6.sspro.p2.T.tdiag.grid2-d3.boot-resid.ci95.0\endcsname{-0.028}
\expandafter\def\csname cnt@rev.desc.d6.sspro.p2.T.tdiag.grid2-d3.boot-resid.ci95.0\endcsname{0}
\expandafter\def\csname pct@rev.desc.d6.sspro.p2.T.tdiag.grid2-d3.boot-resid.ci95.0\endcsname{$-$2.8}
\expandafter\def\csname ppt@rev.desc.d6.sspro.p2.T.tdiag.grid2-d3.boot-resid.ci95.0\endcsname{$-$2.8}
\expandafter\def\csname val@rev.desc.d6.sspro.p2.T.tdiag.grid2-d3.boot-resid.ci95.1\endcsname{0.052}
\expandafter\def\csname cnt@rev.desc.d6.sspro.p2.T.tdiag.grid2-d3.boot-resid.ci95.1\endcsname{0}
\expandafter\def\csname pct@rev.desc.d6.sspro.p2.T.tdiag.grid2-d3.boot-resid.ci95.1\endcsname{5.2}
\expandafter\def\csname ppt@rev.desc.d6.sspro.p2.T.tdiag.grid2-d3.boot-resid.ci95.1\endcsname{$+$5.2}
\expandafter\def\csname val@rev.desc.d6.sspro.p2.T.tdiag.grid2-d3.boot-resid.n-apps\endcsname{12}
\expandafter\def\csname cnt@rev.desc.d6.sspro.p2.T.tdiag.grid2-d3.boot-resid.n-apps\endcsname{12}
\expandafter\def\csname pct@rev.desc.d6.sspro.p2.T.tdiag.grid2-d3.boot-resid.n-apps\endcsname{1200.0}
\expandafter\def\csname ppt@rev.desc.d6.sspro.p2.T.tdiag.grid2-d3.boot-resid.n-apps\endcsname{$+$1200.0}
\expandafter\def\csname val@rev.desc.d6.sspro.p2.T.tdiag.grid2-d3.boot-resid.n-boot-valid\endcsname{4000}
\expandafter\def\csname cnt@rev.desc.d6.sspro.p2.T.tdiag.grid2-d3.boot-resid.n-boot-valid\endcsname{4{,}000}
\expandafter\def\csname pct@rev.desc.d6.sspro.p2.T.tdiag.grid2-d3.boot-resid.n-boot-valid\endcsname{400000.0}
\expandafter\def\csname ppt@rev.desc.d6.sspro.p2.T.tdiag.grid2-d3.boot-resid.n-boot-valid\endcsname{$+$400000.0}
\expandafter\def\csname val@rev.desc.d6.sspro.p2.T.tdiag.grid2-d3.boot-resid.value\endcsname{0.013}
\expandafter\def\csname cnt@rev.desc.d6.sspro.p2.T.tdiag.grid2-d3.boot-resid.value\endcsname{0}
\expandafter\def\csname pct@rev.desc.d6.sspro.p2.T.tdiag.grid2-d3.boot-resid.value\endcsname{1.3}
\expandafter\def\csname ppt@rev.desc.d6.sspro.p2.T.tdiag.grid2-d3.boot-resid.value\endcsname{$+$1.3}
\expandafter\def\csname val@rev.desc.d6.sspro.p2.T.tdiag.grid2-d3.boot-share.ci95.0\endcsname{0.740}
\expandafter\def\csname cnt@rev.desc.d6.sspro.p2.T.tdiag.grid2-d3.boot-share.ci95.0\endcsname{1}
\expandafter\def\csname pct@rev.desc.d6.sspro.p2.T.tdiag.grid2-d3.boot-share.ci95.0\endcsname{74.0}
\expandafter\def\csname ppt@rev.desc.d6.sspro.p2.T.tdiag.grid2-d3.boot-share.ci95.0\endcsname{$+$74.0}
\expandafter\def\csname val@rev.desc.d6.sspro.p2.T.tdiag.grid2-d3.boot-share.ci95.1\endcsname{0.896}
\expandafter\def\csname cnt@rev.desc.d6.sspro.p2.T.tdiag.grid2-d3.boot-share.ci95.1\endcsname{1}
\expandafter\def\csname pct@rev.desc.d6.sspro.p2.T.tdiag.grid2-d3.boot-share.ci95.1\endcsname{89.6}
\expandafter\def\csname ppt@rev.desc.d6.sspro.p2.T.tdiag.grid2-d3.boot-share.ci95.1\endcsname{$+$89.6}
\expandafter\def\csname val@rev.desc.d6.sspro.p2.T.tdiag.grid2-d3.boot-share.n-apps\endcsname{12}
\expandafter\def\csname cnt@rev.desc.d6.sspro.p2.T.tdiag.grid2-d3.boot-share.n-apps\endcsname{12}
\expandafter\def\csname pct@rev.desc.d6.sspro.p2.T.tdiag.grid2-d3.boot-share.n-apps\endcsname{1200.0}
\expandafter\def\csname ppt@rev.desc.d6.sspro.p2.T.tdiag.grid2-d3.boot-share.n-apps\endcsname{$+$1200.0}
\expandafter\def\csname val@rev.desc.d6.sspro.p2.T.tdiag.grid2-d3.boot-share.n-boot-valid\endcsname{4000}
\expandafter\def\csname cnt@rev.desc.d6.sspro.p2.T.tdiag.grid2-d3.boot-share.n-boot-valid\endcsname{4{,}000}
\expandafter\def\csname pct@rev.desc.d6.sspro.p2.T.tdiag.grid2-d3.boot-share.n-boot-valid\endcsname{400000.0}
\expandafter\def\csname ppt@rev.desc.d6.sspro.p2.T.tdiag.grid2-d3.boot-share.n-boot-valid\endcsname{$+$400000.0}
\expandafter\def\csname val@rev.desc.d6.sspro.p2.T.tdiag.grid2-d3.boot-share.value\endcsname{0.823}
\expandafter\def\csname cnt@rev.desc.d6.sspro.p2.T.tdiag.grid2-d3.boot-share.value\endcsname{1}
\expandafter\def\csname pct@rev.desc.d6.sspro.p2.T.tdiag.grid2-d3.boot-share.value\endcsname{82.3}
\expandafter\def\csname ppt@rev.desc.d6.sspro.p2.T.tdiag.grid2-d3.boot-share.value\endcsname{$+$82.3}
\expandafter\def\csname val@rev.desc.d6.sspro.p2.T.tdiag.grid2-d3.n\endcsname{96}
\expandafter\def\csname cnt@rev.desc.d6.sspro.p2.T.tdiag.grid2-d3.n\endcsname{96}
\expandafter\def\csname pct@rev.desc.d6.sspro.p2.T.tdiag.grid2-d3.n\endcsname{9600.0}
\expandafter\def\csname ppt@rev.desc.d6.sspro.p2.T.tdiag.grid2-d3.n\endcsname{$+$9600.0}
\expandafter\def\csname val@rev.desc.d6.sspro.p2.T.tdiag.grid2-d3.resid-agree-stable\endcsname{0.013}
\expandafter\def\csname cnt@rev.desc.d6.sspro.p2.T.tdiag.grid2-d3.resid-agree-stable\endcsname{0}
\expandafter\def\csname pct@rev.desc.d6.sspro.p2.T.tdiag.grid2-d3.resid-agree-stable\endcsname{1.3}
\expandafter\def\csname ppt@rev.desc.d6.sspro.p2.T.tdiag.grid2-d3.resid-agree-stable\endcsname{$+$1.3}
\expandafter\def\csname val@rev.desc.d6.sspro.p2.T.tdiag.grid2-d3.resid-disagree\endcsname{0.000}
\expandafter\def\csname cnt@rev.desc.d6.sspro.p2.T.tdiag.grid2-d3.resid-disagree\endcsname{0}
\expandafter\def\csname pct@rev.desc.d6.sspro.p2.T.tdiag.grid2-d3.resid-disagree\endcsname{0.0}
\expandafter\def\csname ppt@rev.desc.d6.sspro.p2.T.tdiag.grid2-d3.resid-disagree\endcsname{0.0}
\expandafter\def\csname val@rev.desc.d6.sspro.p2.T.tdiag.grid2-d3.share-agree-stable\endcsname{0.823}
\expandafter\def\csname cnt@rev.desc.d6.sspro.p2.T.tdiag.grid2-d3.share-agree-stable\endcsname{1}
\expandafter\def\csname pct@rev.desc.d6.sspro.p2.T.tdiag.grid2-d3.share-agree-stable\endcsname{82.3}
\expandafter\def\csname ppt@rev.desc.d6.sspro.p2.T.tdiag.grid2-d3.share-agree-stable\endcsname{$+$82.3}
\expandafter\def\csname val@rev.desc.d6.sspro.p2.T.tdiag.zoomclick-d3.N-agree-stable\endcsname{81}
\expandafter\def\csname cnt@rev.desc.d6.sspro.p2.T.tdiag.zoomclick-d3.N-agree-stable\endcsname{81}
\expandafter\def\csname pct@rev.desc.d6.sspro.p2.T.tdiag.zoomclick-d3.N-agree-stable\endcsname{8100.0}
\expandafter\def\csname ppt@rev.desc.d6.sspro.p2.T.tdiag.zoomclick-d3.N-agree-stable\endcsname{$+$8100.0}
\expandafter\def\csname val@rev.desc.d6.sspro.p2.T.tdiag.zoomclick-d3.N-disagree\endcsname{13}
\expandafter\def\csname cnt@rev.desc.d6.sspro.p2.T.tdiag.zoomclick-d3.N-disagree\endcsname{13}
\expandafter\def\csname pct@rev.desc.d6.sspro.p2.T.tdiag.zoomclick-d3.N-disagree\endcsname{1300.0}
\expandafter\def\csname ppt@rev.desc.d6.sspro.p2.T.tdiag.zoomclick-d3.N-disagree\endcsname{$+$1300.0}
\expandafter\def\csname val@rev.desc.d6.sspro.p2.T.tdiag.zoomclick-d3.boot-resid.ci95.0\endcsname{0.000}
\expandafter\def\csname cnt@rev.desc.d6.sspro.p2.T.tdiag.zoomclick-d3.boot-resid.ci95.0\endcsname{0}
\expandafter\def\csname pct@rev.desc.d6.sspro.p2.T.tdiag.zoomclick-d3.boot-resid.ci95.0\endcsname{0.0}
\expandafter\def\csname ppt@rev.desc.d6.sspro.p2.T.tdiag.zoomclick-d3.boot-resid.ci95.0\endcsname{0.0}
\expandafter\def\csname val@rev.desc.d6.sspro.p2.T.tdiag.zoomclick-d3.boot-resid.ci95.1\endcsname{0.000}
\expandafter\def\csname cnt@rev.desc.d6.sspro.p2.T.tdiag.zoomclick-d3.boot-resid.ci95.1\endcsname{0}
\expandafter\def\csname pct@rev.desc.d6.sspro.p2.T.tdiag.zoomclick-d3.boot-resid.ci95.1\endcsname{0.0}
\expandafter\def\csname ppt@rev.desc.d6.sspro.p2.T.tdiag.zoomclick-d3.boot-resid.ci95.1\endcsname{0.0}
\expandafter\def\csname val@rev.desc.d6.sspro.p2.T.tdiag.zoomclick-d3.boot-resid.n-apps\endcsname{12}
\expandafter\def\csname cnt@rev.desc.d6.sspro.p2.T.tdiag.zoomclick-d3.boot-resid.n-apps\endcsname{12}
\expandafter\def\csname pct@rev.desc.d6.sspro.p2.T.tdiag.zoomclick-d3.boot-resid.n-apps\endcsname{1200.0}
\expandafter\def\csname ppt@rev.desc.d6.sspro.p2.T.tdiag.zoomclick-d3.boot-resid.n-apps\endcsname{$+$1200.0}
\expandafter\def\csname val@rev.desc.d6.sspro.p2.T.tdiag.zoomclick-d3.boot-resid.n-boot-valid\endcsname{4000}
\expandafter\def\csname cnt@rev.desc.d6.sspro.p2.T.tdiag.zoomclick-d3.boot-resid.n-boot-valid\endcsname{4{,}000}
\expandafter\def\csname pct@rev.desc.d6.sspro.p2.T.tdiag.zoomclick-d3.boot-resid.n-boot-valid\endcsname{400000.0}
\expandafter\def\csname ppt@rev.desc.d6.sspro.p2.T.tdiag.zoomclick-d3.boot-resid.n-boot-valid\endcsname{$+$400000.0}
\expandafter\def\csname val@rev.desc.d6.sspro.p2.T.tdiag.zoomclick-d3.boot-resid.value\endcsname{0.000}
\expandafter\def\csname cnt@rev.desc.d6.sspro.p2.T.tdiag.zoomclick-d3.boot-resid.value\endcsname{0}
\expandafter\def\csname pct@rev.desc.d6.sspro.p2.T.tdiag.zoomclick-d3.boot-resid.value\endcsname{0.0}
\expandafter\def\csname ppt@rev.desc.d6.sspro.p2.T.tdiag.zoomclick-d3.boot-resid.value\endcsname{0.0}
\expandafter\def\csname val@rev.desc.d6.sspro.p2.T.tdiag.zoomclick-d3.boot-share.ci95.0\endcsname{0.760}
\expandafter\def\csname cnt@rev.desc.d6.sspro.p2.T.tdiag.zoomclick-d3.boot-share.ci95.0\endcsname{1}
\expandafter\def\csname pct@rev.desc.d6.sspro.p2.T.tdiag.zoomclick-d3.boot-share.ci95.0\endcsname{76.0}
\expandafter\def\csname ppt@rev.desc.d6.sspro.p2.T.tdiag.zoomclick-d3.boot-share.ci95.0\endcsname{$+$76.0}
\expandafter\def\csname val@rev.desc.d6.sspro.p2.T.tdiag.zoomclick-d3.boot-share.ci95.1\endcsname{0.906}
\expandafter\def\csname cnt@rev.desc.d6.sspro.p2.T.tdiag.zoomclick-d3.boot-share.ci95.1\endcsname{1}
\expandafter\def\csname pct@rev.desc.d6.sspro.p2.T.tdiag.zoomclick-d3.boot-share.ci95.1\endcsname{90.6}
\expandafter\def\csname ppt@rev.desc.d6.sspro.p2.T.tdiag.zoomclick-d3.boot-share.ci95.1\endcsname{$+$90.6}
\expandafter\def\csname val@rev.desc.d6.sspro.p2.T.tdiag.zoomclick-d3.boot-share.n-apps\endcsname{12}
\expandafter\def\csname cnt@rev.desc.d6.sspro.p2.T.tdiag.zoomclick-d3.boot-share.n-apps\endcsname{12}
\expandafter\def\csname pct@rev.desc.d6.sspro.p2.T.tdiag.zoomclick-d3.boot-share.n-apps\endcsname{1200.0}
\expandafter\def\csname ppt@rev.desc.d6.sspro.p2.T.tdiag.zoomclick-d3.boot-share.n-apps\endcsname{$+$1200.0}
\expandafter\def\csname val@rev.desc.d6.sspro.p2.T.tdiag.zoomclick-d3.boot-share.n-boot-valid\endcsname{4000}
\expandafter\def\csname cnt@rev.desc.d6.sspro.p2.T.tdiag.zoomclick-d3.boot-share.n-boot-valid\endcsname{4{,}000}
\expandafter\def\csname pct@rev.desc.d6.sspro.p2.T.tdiag.zoomclick-d3.boot-share.n-boot-valid\endcsname{400000.0}
\expandafter\def\csname ppt@rev.desc.d6.sspro.p2.T.tdiag.zoomclick-d3.boot-share.n-boot-valid\endcsname{$+$400000.0}
\expandafter\def\csname val@rev.desc.d6.sspro.p2.T.tdiag.zoomclick-d3.boot-share.value\endcsname{0.844}
\expandafter\def\csname cnt@rev.desc.d6.sspro.p2.T.tdiag.zoomclick-d3.boot-share.value\endcsname{1}
\expandafter\def\csname pct@rev.desc.d6.sspro.p2.T.tdiag.zoomclick-d3.boot-share.value\endcsname{84.4}
\expandafter\def\csname ppt@rev.desc.d6.sspro.p2.T.tdiag.zoomclick-d3.boot-share.value\endcsname{$+$84.4}
\expandafter\def\csname val@rev.desc.d6.sspro.p2.T.tdiag.zoomclick-d3.n\endcsname{96}
\expandafter\def\csname cnt@rev.desc.d6.sspro.p2.T.tdiag.zoomclick-d3.n\endcsname{96}
\expandafter\def\csname pct@rev.desc.d6.sspro.p2.T.tdiag.zoomclick-d3.n\endcsname{9600.0}
\expandafter\def\csname ppt@rev.desc.d6.sspro.p2.T.tdiag.zoomclick-d3.n\endcsname{$+$9600.0}
\expandafter\def\csname val@rev.desc.d6.sspro.p2.T.tdiag.zoomclick-d3.resid-agree-stable\endcsname{0.000}
\expandafter\def\csname cnt@rev.desc.d6.sspro.p2.T.tdiag.zoomclick-d3.resid-agree-stable\endcsname{0}
\expandafter\def\csname pct@rev.desc.d6.sspro.p2.T.tdiag.zoomclick-d3.resid-agree-stable\endcsname{0.0}
\expandafter\def\csname ppt@rev.desc.d6.sspro.p2.T.tdiag.zoomclick-d3.resid-agree-stable\endcsname{0.0}
\expandafter\def\csname val@rev.desc.d6.sspro.p2.T.tdiag.zoomclick-d3.resid-disagree\endcsname{0.077}
\expandafter\def\csname cnt@rev.desc.d6.sspro.p2.T.tdiag.zoomclick-d3.resid-disagree\endcsname{0}
\expandafter\def\csname pct@rev.desc.d6.sspro.p2.T.tdiag.zoomclick-d3.resid-disagree\endcsname{7.7}
\expandafter\def\csname ppt@rev.desc.d6.sspro.p2.T.tdiag.zoomclick-d3.resid-disagree\endcsname{$+$7.7}
\expandafter\def\csname val@rev.desc.d6.sspro.p2.T.tdiag.zoomclick-d3.share-agree-stable\endcsname{0.844}
\expandafter\def\csname cnt@rev.desc.d6.sspro.p2.T.tdiag.zoomclick-d3.share-agree-stable\endcsname{1}
\expandafter\def\csname pct@rev.desc.d6.sspro.p2.T.tdiag.zoomclick-d3.share-agree-stable\endcsname{84.4}
\expandafter\def\csname ppt@rev.desc.d6.sspro.p2.T.tdiag.zoomclick-d3.share-agree-stable\endcsname{$+$84.4}
\expandafter\def\csname val@rev.desc.d6.uiv.p1.C.box.catalog-pooled.N-agree-stable\endcsname{223}
\expandafter\def\csname cnt@rev.desc.d6.uiv.p1.C.box.catalog-pooled.N-agree-stable\endcsname{223}
\expandafter\def\csname pct@rev.desc.d6.uiv.p1.C.box.catalog-pooled.N-agree-stable\endcsname{22300.0}
\expandafter\def\csname ppt@rev.desc.d6.uiv.p1.C.box.catalog-pooled.N-agree-stable\endcsname{$+$22300.0}
\expandafter\def\csname val@rev.desc.d6.uiv.p1.C.box.catalog-pooled.N-disagree\endcsname{130}
\expandafter\def\csname cnt@rev.desc.d6.uiv.p1.C.box.catalog-pooled.N-disagree\endcsname{130}
\expandafter\def\csname pct@rev.desc.d6.uiv.p1.C.box.catalog-pooled.N-disagree\endcsname{13000.0}
\expandafter\def\csname ppt@rev.desc.d6.uiv.p1.C.box.catalog-pooled.N-disagree\endcsname{$+$13000.0}
\expandafter\def\csname val@rev.desc.d6.uiv.p1.C.box.catalog-pooled.boot-resid.ci95.0\endcsname{-0.013}
\expandafter\def\csname cnt@rev.desc.d6.uiv.p1.C.box.catalog-pooled.boot-resid.ci95.0\endcsname{0}
\expandafter\def\csname pct@rev.desc.d6.uiv.p1.C.box.catalog-pooled.boot-resid.ci95.0\endcsname{$-$1.3}
\expandafter\def\csname ppt@rev.desc.d6.uiv.p1.C.box.catalog-pooled.boot-resid.ci95.0\endcsname{$-$1.3}
\expandafter\def\csname val@rev.desc.d6.uiv.p1.C.box.catalog-pooled.boot-resid.ci95.1\endcsname{0.000}
\expandafter\def\csname cnt@rev.desc.d6.uiv.p1.C.box.catalog-pooled.boot-resid.ci95.1\endcsname{0}
\expandafter\def\csname pct@rev.desc.d6.uiv.p1.C.box.catalog-pooled.boot-resid.ci95.1\endcsname{0.0}
\expandafter\def\csname ppt@rev.desc.d6.uiv.p1.C.box.catalog-pooled.boot-resid.ci95.1\endcsname{0.0}
\expandafter\def\csname val@rev.desc.d6.uiv.p1.C.box.catalog-pooled.boot-resid.n-apps\endcsname{12}
\expandafter\def\csname cnt@rev.desc.d6.uiv.p1.C.box.catalog-pooled.boot-resid.n-apps\endcsname{12}
\expandafter\def\csname pct@rev.desc.d6.uiv.p1.C.box.catalog-pooled.boot-resid.n-apps\endcsname{1200.0}
\expandafter\def\csname ppt@rev.desc.d6.uiv.p1.C.box.catalog-pooled.boot-resid.n-apps\endcsname{$+$1200.0}
\expandafter\def\csname val@rev.desc.d6.uiv.p1.C.box.catalog-pooled.boot-resid.n-boot-valid\endcsname{4000}
\expandafter\def\csname cnt@rev.desc.d6.uiv.p1.C.box.catalog-pooled.boot-resid.n-boot-valid\endcsname{4{,}000}
\expandafter\def\csname pct@rev.desc.d6.uiv.p1.C.box.catalog-pooled.boot-resid.n-boot-valid\endcsname{400000.0}
\expandafter\def\csname ppt@rev.desc.d6.uiv.p1.C.box.catalog-pooled.boot-resid.n-boot-valid\endcsname{$+$400000.0}
\expandafter\def\csname val@rev.desc.d6.uiv.p1.C.box.catalog-pooled.boot-resid.value\endcsname{-0.004}
\expandafter\def\csname cnt@rev.desc.d6.uiv.p1.C.box.catalog-pooled.boot-resid.value\endcsname{0}
\expandafter\def\csname pct@rev.desc.d6.uiv.p1.C.box.catalog-pooled.boot-resid.value\endcsname{$-$0.4}
\expandafter\def\csname ppt@rev.desc.d6.uiv.p1.C.box.catalog-pooled.boot-resid.value\endcsname{$-$0.4}
\expandafter\def\csname val@rev.desc.d6.uiv.p1.C.box.catalog-pooled.boot-share.ci95.0\endcsname{0.518}
\expandafter\def\csname cnt@rev.desc.d6.uiv.p1.C.box.catalog-pooled.boot-share.ci95.0\endcsname{1}
\expandafter\def\csname pct@rev.desc.d6.uiv.p1.C.box.catalog-pooled.boot-share.ci95.0\endcsname{51.8}
\expandafter\def\csname ppt@rev.desc.d6.uiv.p1.C.box.catalog-pooled.boot-share.ci95.0\endcsname{$+$51.8}
\expandafter\def\csname val@rev.desc.d6.uiv.p1.C.box.catalog-pooled.boot-share.ci95.1\endcsname{0.643}
\expandafter\def\csname cnt@rev.desc.d6.uiv.p1.C.box.catalog-pooled.boot-share.ci95.1\endcsname{1}
\expandafter\def\csname pct@rev.desc.d6.uiv.p1.C.box.catalog-pooled.boot-share.ci95.1\endcsname{64.3}
\expandafter\def\csname ppt@rev.desc.d6.uiv.p1.C.box.catalog-pooled.boot-share.ci95.1\endcsname{$+$64.3}
\expandafter\def\csname val@rev.desc.d6.uiv.p1.C.box.catalog-pooled.boot-share.n-apps\endcsname{12}
\expandafter\def\csname cnt@rev.desc.d6.uiv.p1.C.box.catalog-pooled.boot-share.n-apps\endcsname{12}
\expandafter\def\csname pct@rev.desc.d6.uiv.p1.C.box.catalog-pooled.boot-share.n-apps\endcsname{1200.0}
\expandafter\def\csname ppt@rev.desc.d6.uiv.p1.C.box.catalog-pooled.boot-share.n-apps\endcsname{$+$1200.0}
\expandafter\def\csname val@rev.desc.d6.uiv.p1.C.box.catalog-pooled.boot-share.n-boot-valid\endcsname{4000}
\expandafter\def\csname cnt@rev.desc.d6.uiv.p1.C.box.catalog-pooled.boot-share.n-boot-valid\endcsname{4{,}000}
\expandafter\def\csname pct@rev.desc.d6.uiv.p1.C.box.catalog-pooled.boot-share.n-boot-valid\endcsname{400000.0}
\expandafter\def\csname ppt@rev.desc.d6.uiv.p1.C.box.catalog-pooled.boot-share.n-boot-valid\endcsname{$+$400000.0}
\expandafter\def\csname val@rev.desc.d6.uiv.p1.C.box.catalog-pooled.boot-share.value\endcsname{0.581}
\expandafter\def\csname cnt@rev.desc.d6.uiv.p1.C.box.catalog-pooled.boot-share.value\endcsname{1}
\expandafter\def\csname pct@rev.desc.d6.uiv.p1.C.box.catalog-pooled.boot-share.value\endcsname{58.1}
\expandafter\def\csname ppt@rev.desc.d6.uiv.p1.C.box.catalog-pooled.boot-share.value\endcsname{$+$58.1}
\expandafter\def\csname val@rev.desc.d6.uiv.p1.C.box.catalog-pooled.gap\endcsname{0.220}
\expandafter\def\csname cnt@rev.desc.d6.uiv.p1.C.box.catalog-pooled.gap\endcsname{0}
\expandafter\def\csname pct@rev.desc.d6.uiv.p1.C.box.catalog-pooled.gap\endcsname{22.0}
\expandafter\def\csname ppt@rev.desc.d6.uiv.p1.C.box.catalog-pooled.gap\endcsname{$+$22.0}
\expandafter\def\csname val@rev.desc.d6.uiv.p1.C.box.catalog-pooled.n\endcsname{384}
\expandafter\def\csname cnt@rev.desc.d6.uiv.p1.C.box.catalog-pooled.n\endcsname{384}
\expandafter\def\csname pct@rev.desc.d6.uiv.p1.C.box.catalog-pooled.n\endcsname{38400.0}
\expandafter\def\csname ppt@rev.desc.d6.uiv.p1.C.box.catalog-pooled.n\endcsname{$+$38400.0}
\expandafter\def\csname val@rev.desc.d6.uiv.p1.C.box.catalog-pooled.resid-agree-stable\endcsname{-0.004}
\expandafter\def\csname cnt@rev.desc.d6.uiv.p1.C.box.catalog-pooled.resid-agree-stable\endcsname{0}
\expandafter\def\csname pct@rev.desc.d6.uiv.p1.C.box.catalog-pooled.resid-agree-stable\endcsname{$-$0.4}
\expandafter\def\csname ppt@rev.desc.d6.uiv.p1.C.box.catalog-pooled.resid-agree-stable\endcsname{$-$0.4}
\expandafter\def\csname val@rev.desc.d6.uiv.p1.C.box.catalog-pooled.resid-disagree\endcsname{0.215}
\expandafter\def\csname cnt@rev.desc.d6.uiv.p1.C.box.catalog-pooled.resid-disagree\endcsname{0}
\expandafter\def\csname pct@rev.desc.d6.uiv.p1.C.box.catalog-pooled.resid-disagree\endcsname{21.5}
\expandafter\def\csname ppt@rev.desc.d6.uiv.p1.C.box.catalog-pooled.resid-disagree\endcsname{$+$21.5}
\expandafter\def\csname val@rev.desc.d6.uiv.p1.C.box.catalog-pooled.share-agree-stable\endcsname{0.581}
\expandafter\def\csname cnt@rev.desc.d6.uiv.p1.C.box.catalog-pooled.share-agree-stable\endcsname{1}
\expandafter\def\csname pct@rev.desc.d6.uiv.p1.C.box.catalog-pooled.share-agree-stable\endcsname{58.1}
\expandafter\def\csname ppt@rev.desc.d6.uiv.p1.C.box.catalog-pooled.share-agree-stable\endcsname{$+$58.1}
\expandafter\def\csname val@rev.desc.d6.uiv.p1.C.box.center-d1.N-agree-stable\endcsname{58}
\expandafter\def\csname cnt@rev.desc.d6.uiv.p1.C.box.center-d1.N-agree-stable\endcsname{58}
\expandafter\def\csname pct@rev.desc.d6.uiv.p1.C.box.center-d1.N-agree-stable\endcsname{5800.0}
\expandafter\def\csname ppt@rev.desc.d6.uiv.p1.C.box.center-d1.N-agree-stable\endcsname{$+$5800.0}
\expandafter\def\csname val@rev.desc.d6.uiv.p1.C.box.center-d1.N-disagree\endcsname{32}
\expandafter\def\csname cnt@rev.desc.d6.uiv.p1.C.box.center-d1.N-disagree\endcsname{32}
\expandafter\def\csname pct@rev.desc.d6.uiv.p1.C.box.center-d1.N-disagree\endcsname{3200.0}
\expandafter\def\csname ppt@rev.desc.d6.uiv.p1.C.box.center-d1.N-disagree\endcsname{$+$3200.0}
\expandafter\def\csname val@rev.desc.d6.uiv.p1.C.box.center-d1.boot-resid.ci95.0\endcsname{0.000}
\expandafter\def\csname cnt@rev.desc.d6.uiv.p1.C.box.center-d1.boot-resid.ci95.0\endcsname{0}
\expandafter\def\csname pct@rev.desc.d6.uiv.p1.C.box.center-d1.boot-resid.ci95.0\endcsname{0.0}
\expandafter\def\csname ppt@rev.desc.d6.uiv.p1.C.box.center-d1.boot-resid.ci95.0\endcsname{0.0}
\expandafter\def\csname val@rev.desc.d6.uiv.p1.C.box.center-d1.boot-resid.ci95.1\endcsname{0.000}
\expandafter\def\csname cnt@rev.desc.d6.uiv.p1.C.box.center-d1.boot-resid.ci95.1\endcsname{0}
\expandafter\def\csname pct@rev.desc.d6.uiv.p1.C.box.center-d1.boot-resid.ci95.1\endcsname{0.0}
\expandafter\def\csname ppt@rev.desc.d6.uiv.p1.C.box.center-d1.boot-resid.ci95.1\endcsname{0.0}
\expandafter\def\csname val@rev.desc.d6.uiv.p1.C.box.center-d1.boot-resid.n-apps\endcsname{12}
\expandafter\def\csname cnt@rev.desc.d6.uiv.p1.C.box.center-d1.boot-resid.n-apps\endcsname{12}
\expandafter\def\csname pct@rev.desc.d6.uiv.p1.C.box.center-d1.boot-resid.n-apps\endcsname{1200.0}
\expandafter\def\csname ppt@rev.desc.d6.uiv.p1.C.box.center-d1.boot-resid.n-apps\endcsname{$+$1200.0}
\expandafter\def\csname val@rev.desc.d6.uiv.p1.C.box.center-d1.boot-resid.n-boot-valid\endcsname{4000}
\expandafter\def\csname cnt@rev.desc.d6.uiv.p1.C.box.center-d1.boot-resid.n-boot-valid\endcsname{4{,}000}
\expandafter\def\csname pct@rev.desc.d6.uiv.p1.C.box.center-d1.boot-resid.n-boot-valid\endcsname{400000.0}
\expandafter\def\csname ppt@rev.desc.d6.uiv.p1.C.box.center-d1.boot-resid.n-boot-valid\endcsname{$+$400000.0}
\expandafter\def\csname val@rev.desc.d6.uiv.p1.C.box.center-d1.boot-resid.value\endcsname{0.000}
\expandafter\def\csname cnt@rev.desc.d6.uiv.p1.C.box.center-d1.boot-resid.value\endcsname{0}
\expandafter\def\csname pct@rev.desc.d6.uiv.p1.C.box.center-d1.boot-resid.value\endcsname{0.0}
\expandafter\def\csname ppt@rev.desc.d6.uiv.p1.C.box.center-d1.boot-resid.value\endcsname{0.0}
\expandafter\def\csname val@rev.desc.d6.uiv.p1.C.box.center-d1.boot-share.ci95.0\endcsname{0.552}
\expandafter\def\csname cnt@rev.desc.d6.uiv.p1.C.box.center-d1.boot-share.ci95.0\endcsname{1}
\expandafter\def\csname pct@rev.desc.d6.uiv.p1.C.box.center-d1.boot-share.ci95.0\endcsname{55.2}
\expandafter\def\csname ppt@rev.desc.d6.uiv.p1.C.box.center-d1.boot-share.ci95.0\endcsname{$+$55.2}
\expandafter\def\csname val@rev.desc.d6.uiv.p1.C.box.center-d1.boot-share.ci95.1\endcsname{0.656}
\expandafter\def\csname cnt@rev.desc.d6.uiv.p1.C.box.center-d1.boot-share.ci95.1\endcsname{1}
\expandafter\def\csname pct@rev.desc.d6.uiv.p1.C.box.center-d1.boot-share.ci95.1\endcsname{65.6}
\expandafter\def\csname ppt@rev.desc.d6.uiv.p1.C.box.center-d1.boot-share.ci95.1\endcsname{$+$65.6}
\expandafter\def\csname val@rev.desc.d6.uiv.p1.C.box.center-d1.boot-share.n-apps\endcsname{12}
\expandafter\def\csname cnt@rev.desc.d6.uiv.p1.C.box.center-d1.boot-share.n-apps\endcsname{12}
\expandafter\def\csname pct@rev.desc.d6.uiv.p1.C.box.center-d1.boot-share.n-apps\endcsname{1200.0}
\expandafter\def\csname ppt@rev.desc.d6.uiv.p1.C.box.center-d1.boot-share.n-apps\endcsname{$+$1200.0}
\expandafter\def\csname val@rev.desc.d6.uiv.p1.C.box.center-d1.boot-share.n-boot-valid\endcsname{4000}
\expandafter\def\csname cnt@rev.desc.d6.uiv.p1.C.box.center-d1.boot-share.n-boot-valid\endcsname{4{,}000}
\expandafter\def\csname pct@rev.desc.d6.uiv.p1.C.box.center-d1.boot-share.n-boot-valid\endcsname{400000.0}
\expandafter\def\csname ppt@rev.desc.d6.uiv.p1.C.box.center-d1.boot-share.n-boot-valid\endcsname{$+$400000.0}
\expandafter\def\csname val@rev.desc.d6.uiv.p1.C.box.center-d1.boot-share.value\endcsname{0.604}
\expandafter\def\csname cnt@rev.desc.d6.uiv.p1.C.box.center-d1.boot-share.value\endcsname{1}
\expandafter\def\csname pct@rev.desc.d6.uiv.p1.C.box.center-d1.boot-share.value\endcsname{60.4}
\expandafter\def\csname ppt@rev.desc.d6.uiv.p1.C.box.center-d1.boot-share.value\endcsname{$+$60.4}
\expandafter\def\csname val@rev.desc.d6.uiv.p1.C.box.center-d1.n\endcsname{96}
\expandafter\def\csname cnt@rev.desc.d6.uiv.p1.C.box.center-d1.n\endcsname{96}
\expandafter\def\csname pct@rev.desc.d6.uiv.p1.C.box.center-d1.n\endcsname{9600.0}
\expandafter\def\csname ppt@rev.desc.d6.uiv.p1.C.box.center-d1.n\endcsname{$+$9600.0}
\expandafter\def\csname val@rev.desc.d6.uiv.p1.C.box.center-d1.resid-agree-stable\endcsname{0.000}
\expandafter\def\csname cnt@rev.desc.d6.uiv.p1.C.box.center-d1.resid-agree-stable\endcsname{0}
\expandafter\def\csname pct@rev.desc.d6.uiv.p1.C.box.center-d1.resid-agree-stable\endcsname{0.0}
\expandafter\def\csname ppt@rev.desc.d6.uiv.p1.C.box.center-d1.resid-agree-stable\endcsname{0.0}
\expandafter\def\csname val@rev.desc.d6.uiv.p1.C.box.center-d1.resid-disagree\endcsname{0.219}
\expandafter\def\csname cnt@rev.desc.d6.uiv.p1.C.box.center-d1.resid-disagree\endcsname{0}
\expandafter\def\csname pct@rev.desc.d6.uiv.p1.C.box.center-d1.resid-disagree\endcsname{21.9}
\expandafter\def\csname ppt@rev.desc.d6.uiv.p1.C.box.center-d1.resid-disagree\endcsname{$+$21.9}
\expandafter\def\csname val@rev.desc.d6.uiv.p1.C.box.center-d1.share-agree-stable\endcsname{0.604}
\expandafter\def\csname cnt@rev.desc.d6.uiv.p1.C.box.center-d1.share-agree-stable\endcsname{1}
\expandafter\def\csname pct@rev.desc.d6.uiv.p1.C.box.center-d1.share-agree-stable\endcsname{60.4}
\expandafter\def\csname ppt@rev.desc.d6.uiv.p1.C.box.center-d1.share-agree-stable\endcsname{$+$60.4}
\expandafter\def\csname val@rev.desc.d6.uiv.p1.C.box.context-d1.N-agree-stable\endcsname{59}
\expandafter\def\csname cnt@rev.desc.d6.uiv.p1.C.box.context-d1.N-agree-stable\endcsname{59}
\expandafter\def\csname pct@rev.desc.d6.uiv.p1.C.box.context-d1.N-agree-stable\endcsname{5900.0}
\expandafter\def\csname ppt@rev.desc.d6.uiv.p1.C.box.context-d1.N-agree-stable\endcsname{$+$5900.0}
\expandafter\def\csname val@rev.desc.d6.uiv.p1.C.box.context-d1.N-disagree\endcsname{28}
\expandafter\def\csname cnt@rev.desc.d6.uiv.p1.C.box.context-d1.N-disagree\endcsname{28}
\expandafter\def\csname pct@rev.desc.d6.uiv.p1.C.box.context-d1.N-disagree\endcsname{2800.0}
\expandafter\def\csname ppt@rev.desc.d6.uiv.p1.C.box.context-d1.N-disagree\endcsname{$+$2800.0}
\expandafter\def\csname val@rev.desc.d6.uiv.p1.C.box.context-d1.boot-resid.ci95.0\endcsname{0.000}
\expandafter\def\csname cnt@rev.desc.d6.uiv.p1.C.box.context-d1.boot-resid.ci95.0\endcsname{0}
\expandafter\def\csname pct@rev.desc.d6.uiv.p1.C.box.context-d1.boot-resid.ci95.0\endcsname{0.0}
\expandafter\def\csname ppt@rev.desc.d6.uiv.p1.C.box.context-d1.boot-resid.ci95.0\endcsname{0.0}
\expandafter\def\csname val@rev.desc.d6.uiv.p1.C.box.context-d1.boot-resid.ci95.1\endcsname{0.000}
\expandafter\def\csname cnt@rev.desc.d6.uiv.p1.C.box.context-d1.boot-resid.ci95.1\endcsname{0}
\expandafter\def\csname pct@rev.desc.d6.uiv.p1.C.box.context-d1.boot-resid.ci95.1\endcsname{0.0}
\expandafter\def\csname ppt@rev.desc.d6.uiv.p1.C.box.context-d1.boot-resid.ci95.1\endcsname{0.0}
\expandafter\def\csname val@rev.desc.d6.uiv.p1.C.box.context-d1.boot-resid.n-apps\endcsname{12}
\expandafter\def\csname cnt@rev.desc.d6.uiv.p1.C.box.context-d1.boot-resid.n-apps\endcsname{12}
\expandafter\def\csname pct@rev.desc.d6.uiv.p1.C.box.context-d1.boot-resid.n-apps\endcsname{1200.0}
\expandafter\def\csname ppt@rev.desc.d6.uiv.p1.C.box.context-d1.boot-resid.n-apps\endcsname{$+$1200.0}
\expandafter\def\csname val@rev.desc.d6.uiv.p1.C.box.context-d1.boot-resid.n-boot-valid\endcsname{4000}
\expandafter\def\csname cnt@rev.desc.d6.uiv.p1.C.box.context-d1.boot-resid.n-boot-valid\endcsname{4{,}000}
\expandafter\def\csname pct@rev.desc.d6.uiv.p1.C.box.context-d1.boot-resid.n-boot-valid\endcsname{400000.0}
\expandafter\def\csname ppt@rev.desc.d6.uiv.p1.C.box.context-d1.boot-resid.n-boot-valid\endcsname{$+$400000.0}
\expandafter\def\csname val@rev.desc.d6.uiv.p1.C.box.context-d1.boot-resid.value\endcsname{0.000}
\expandafter\def\csname cnt@rev.desc.d6.uiv.p1.C.box.context-d1.boot-resid.value\endcsname{0}
\expandafter\def\csname pct@rev.desc.d6.uiv.p1.C.box.context-d1.boot-resid.value\endcsname{0.0}
\expandafter\def\csname ppt@rev.desc.d6.uiv.p1.C.box.context-d1.boot-resid.value\endcsname{0.0}
\expandafter\def\csname val@rev.desc.d6.uiv.p1.C.box.context-d1.boot-share.ci95.0\endcsname{0.542}
\expandafter\def\csname cnt@rev.desc.d6.uiv.p1.C.box.context-d1.boot-share.ci95.0\endcsname{1}
\expandafter\def\csname pct@rev.desc.d6.uiv.p1.C.box.context-d1.boot-share.ci95.0\endcsname{54.2}
\expandafter\def\csname ppt@rev.desc.d6.uiv.p1.C.box.context-d1.boot-share.ci95.0\endcsname{$+$54.2}
\expandafter\def\csname val@rev.desc.d6.uiv.p1.C.box.context-d1.boot-share.ci95.1\endcsname{0.688}
\expandafter\def\csname cnt@rev.desc.d6.uiv.p1.C.box.context-d1.boot-share.ci95.1\endcsname{1}
\expandafter\def\csname pct@rev.desc.d6.uiv.p1.C.box.context-d1.boot-share.ci95.1\endcsname{68.8}
\expandafter\def\csname ppt@rev.desc.d6.uiv.p1.C.box.context-d1.boot-share.ci95.1\endcsname{$+$68.8}
\expandafter\def\csname val@rev.desc.d6.uiv.p1.C.box.context-d1.boot-share.n-apps\endcsname{12}
\expandafter\def\csname cnt@rev.desc.d6.uiv.p1.C.box.context-d1.boot-share.n-apps\endcsname{12}
\expandafter\def\csname pct@rev.desc.d6.uiv.p1.C.box.context-d1.boot-share.n-apps\endcsname{1200.0}
\expandafter\def\csname ppt@rev.desc.d6.uiv.p1.C.box.context-d1.boot-share.n-apps\endcsname{$+$1200.0}
\expandafter\def\csname val@rev.desc.d6.uiv.p1.C.box.context-d1.boot-share.n-boot-valid\endcsname{4000}
\expandafter\def\csname cnt@rev.desc.d6.uiv.p1.C.box.context-d1.boot-share.n-boot-valid\endcsname{4{,}000}
\expandafter\def\csname pct@rev.desc.d6.uiv.p1.C.box.context-d1.boot-share.n-boot-valid\endcsname{400000.0}
\expandafter\def\csname ppt@rev.desc.d6.uiv.p1.C.box.context-d1.boot-share.n-boot-valid\endcsname{$+$400000.0}
\expandafter\def\csname val@rev.desc.d6.uiv.p1.C.box.context-d1.boot-share.value\endcsname{0.615}
\expandafter\def\csname cnt@rev.desc.d6.uiv.p1.C.box.context-d1.boot-share.value\endcsname{1}
\expandafter\def\csname pct@rev.desc.d6.uiv.p1.C.box.context-d1.boot-share.value\endcsname{61.5}
\expandafter\def\csname ppt@rev.desc.d6.uiv.p1.C.box.context-d1.boot-share.value\endcsname{$+$61.5}
\expandafter\def\csname val@rev.desc.d6.uiv.p1.C.box.context-d1.n\endcsname{96}
\expandafter\def\csname cnt@rev.desc.d6.uiv.p1.C.box.context-d1.n\endcsname{96}
\expandafter\def\csname pct@rev.desc.d6.uiv.p1.C.box.context-d1.n\endcsname{9600.0}
\expandafter\def\csname ppt@rev.desc.d6.uiv.p1.C.box.context-d1.n\endcsname{$+$9600.0}
\expandafter\def\csname val@rev.desc.d6.uiv.p1.C.box.context-d1.resid-agree-stable\endcsname{0.000}
\expandafter\def\csname cnt@rev.desc.d6.uiv.p1.C.box.context-d1.resid-agree-stable\endcsname{0}
\expandafter\def\csname pct@rev.desc.d6.uiv.p1.C.box.context-d1.resid-agree-stable\endcsname{0.0}
\expandafter\def\csname ppt@rev.desc.d6.uiv.p1.C.box.context-d1.resid-agree-stable\endcsname{0.0}
\expandafter\def\csname val@rev.desc.d6.uiv.p1.C.box.context-d1.resid-disagree\endcsname{0.036}
\expandafter\def\csname cnt@rev.desc.d6.uiv.p1.C.box.context-d1.resid-disagree\endcsname{0}
\expandafter\def\csname pct@rev.desc.d6.uiv.p1.C.box.context-d1.resid-disagree\endcsname{3.6}
\expandafter\def\csname ppt@rev.desc.d6.uiv.p1.C.box.context-d1.resid-disagree\endcsname{$+$3.6}
\expandafter\def\csname val@rev.desc.d6.uiv.p1.C.box.context-d1.share-agree-stable\endcsname{0.615}
\expandafter\def\csname cnt@rev.desc.d6.uiv.p1.C.box.context-d1.share-agree-stable\endcsname{1}
\expandafter\def\csname pct@rev.desc.d6.uiv.p1.C.box.context-d1.share-agree-stable\endcsname{61.5}
\expandafter\def\csname ppt@rev.desc.d6.uiv.p1.C.box.context-d1.share-agree-stable\endcsname{$+$61.5}
\expandafter\def\csname val@rev.desc.d6.uiv.p1.C.box.dual-d2.N-agree-stable\endcsname{56}
\expandafter\def\csname cnt@rev.desc.d6.uiv.p1.C.box.dual-d2.N-agree-stable\endcsname{56}
\expandafter\def\csname pct@rev.desc.d6.uiv.p1.C.box.dual-d2.N-agree-stable\endcsname{5600.0}
\expandafter\def\csname ppt@rev.desc.d6.uiv.p1.C.box.dual-d2.N-agree-stable\endcsname{$+$5600.0}
\expandafter\def\csname val@rev.desc.d6.uiv.p1.C.box.dual-d2.N-disagree\endcsname{34}
\expandafter\def\csname cnt@rev.desc.d6.uiv.p1.C.box.dual-d2.N-disagree\endcsname{34}
\expandafter\def\csname pct@rev.desc.d6.uiv.p1.C.box.dual-d2.N-disagree\endcsname{3400.0}
\expandafter\def\csname ppt@rev.desc.d6.uiv.p1.C.box.dual-d2.N-disagree\endcsname{$+$3400.0}
\expandafter\def\csname val@rev.desc.d6.uiv.p1.C.box.dual-d2.boot-resid.ci95.0\endcsname{0.000}
\expandafter\def\csname cnt@rev.desc.d6.uiv.p1.C.box.dual-d2.boot-resid.ci95.0\endcsname{0}
\expandafter\def\csname pct@rev.desc.d6.uiv.p1.C.box.dual-d2.boot-resid.ci95.0\endcsname{0.0}
\expandafter\def\csname ppt@rev.desc.d6.uiv.p1.C.box.dual-d2.boot-resid.ci95.0\endcsname{0.0}
\expandafter\def\csname val@rev.desc.d6.uiv.p1.C.box.dual-d2.boot-resid.ci95.1\endcsname{0.000}
\expandafter\def\csname cnt@rev.desc.d6.uiv.p1.C.box.dual-d2.boot-resid.ci95.1\endcsname{0}
\expandafter\def\csname pct@rev.desc.d6.uiv.p1.C.box.dual-d2.boot-resid.ci95.1\endcsname{0.0}
\expandafter\def\csname ppt@rev.desc.d6.uiv.p1.C.box.dual-d2.boot-resid.ci95.1\endcsname{0.0}
\expandafter\def\csname val@rev.desc.d6.uiv.p1.C.box.dual-d2.boot-resid.n-apps\endcsname{12}
\expandafter\def\csname cnt@rev.desc.d6.uiv.p1.C.box.dual-d2.boot-resid.n-apps\endcsname{12}
\expandafter\def\csname pct@rev.desc.d6.uiv.p1.C.box.dual-d2.boot-resid.n-apps\endcsname{1200.0}
\expandafter\def\csname ppt@rev.desc.d6.uiv.p1.C.box.dual-d2.boot-resid.n-apps\endcsname{$+$1200.0}
\expandafter\def\csname val@rev.desc.d6.uiv.p1.C.box.dual-d2.boot-resid.n-boot-valid\endcsname{4000}
\expandafter\def\csname cnt@rev.desc.d6.uiv.p1.C.box.dual-d2.boot-resid.n-boot-valid\endcsname{4{,}000}
\expandafter\def\csname pct@rev.desc.d6.uiv.p1.C.box.dual-d2.boot-resid.n-boot-valid\endcsname{400000.0}
\expandafter\def\csname ppt@rev.desc.d6.uiv.p1.C.box.dual-d2.boot-resid.n-boot-valid\endcsname{$+$400000.0}
\expandafter\def\csname val@rev.desc.d6.uiv.p1.C.box.dual-d2.boot-resid.value\endcsname{0.000}
\expandafter\def\csname cnt@rev.desc.d6.uiv.p1.C.box.dual-d2.boot-resid.value\endcsname{0}
\expandafter\def\csname pct@rev.desc.d6.uiv.p1.C.box.dual-d2.boot-resid.value\endcsname{0.0}
\expandafter\def\csname ppt@rev.desc.d6.uiv.p1.C.box.dual-d2.boot-resid.value\endcsname{0.0}
\expandafter\def\csname val@rev.desc.d6.uiv.p1.C.box.dual-d2.boot-share.ci95.0\endcsname{0.500}
\expandafter\def\csname cnt@rev.desc.d6.uiv.p1.C.box.dual-d2.boot-share.ci95.0\endcsname{0}
\expandafter\def\csname pct@rev.desc.d6.uiv.p1.C.box.dual-d2.boot-share.ci95.0\endcsname{50.0}
\expandafter\def\csname ppt@rev.desc.d6.uiv.p1.C.box.dual-d2.boot-share.ci95.0\endcsname{$+$50.0}
\expandafter\def\csname val@rev.desc.d6.uiv.p1.C.box.dual-d2.boot-share.ci95.1\endcsname{0.667}
\expandafter\def\csname cnt@rev.desc.d6.uiv.p1.C.box.dual-d2.boot-share.ci95.1\endcsname{1}
\expandafter\def\csname pct@rev.desc.d6.uiv.p1.C.box.dual-d2.boot-share.ci95.1\endcsname{66.7}
\expandafter\def\csname ppt@rev.desc.d6.uiv.p1.C.box.dual-d2.boot-share.ci95.1\endcsname{$+$66.7}
\expandafter\def\csname val@rev.desc.d6.uiv.p1.C.box.dual-d2.boot-share.n-apps\endcsname{12}
\expandafter\def\csname cnt@rev.desc.d6.uiv.p1.C.box.dual-d2.boot-share.n-apps\endcsname{12}
\expandafter\def\csname pct@rev.desc.d6.uiv.p1.C.box.dual-d2.boot-share.n-apps\endcsname{1200.0}
\expandafter\def\csname ppt@rev.desc.d6.uiv.p1.C.box.dual-d2.boot-share.n-apps\endcsname{$+$1200.0}
\expandafter\def\csname val@rev.desc.d6.uiv.p1.C.box.dual-d2.boot-share.n-boot-valid\endcsname{4000}
\expandafter\def\csname cnt@rev.desc.d6.uiv.p1.C.box.dual-d2.boot-share.n-boot-valid\endcsname{4{,}000}
\expandafter\def\csname pct@rev.desc.d6.uiv.p1.C.box.dual-d2.boot-share.n-boot-valid\endcsname{400000.0}
\expandafter\def\csname ppt@rev.desc.d6.uiv.p1.C.box.dual-d2.boot-share.n-boot-valid\endcsname{$+$400000.0}
\expandafter\def\csname val@rev.desc.d6.uiv.p1.C.box.dual-d2.boot-share.value\endcsname{0.583}
\expandafter\def\csname cnt@rev.desc.d6.uiv.p1.C.box.dual-d2.boot-share.value\endcsname{1}
\expandafter\def\csname pct@rev.desc.d6.uiv.p1.C.box.dual-d2.boot-share.value\endcsname{58.3}
\expandafter\def\csname ppt@rev.desc.d6.uiv.p1.C.box.dual-d2.boot-share.value\endcsname{$+$58.3}
\expandafter\def\csname val@rev.desc.d6.uiv.p1.C.box.dual-d2.n\endcsname{96}
\expandafter\def\csname cnt@rev.desc.d6.uiv.p1.C.box.dual-d2.n\endcsname{96}
\expandafter\def\csname pct@rev.desc.d6.uiv.p1.C.box.dual-d2.n\endcsname{9600.0}
\expandafter\def\csname ppt@rev.desc.d6.uiv.p1.C.box.dual-d2.n\endcsname{$+$9600.0}
\expandafter\def\csname val@rev.desc.d6.uiv.p1.C.box.dual-d2.resid-agree-stable\endcsname{0.000}
\expandafter\def\csname cnt@rev.desc.d6.uiv.p1.C.box.dual-d2.resid-agree-stable\endcsname{0}
\expandafter\def\csname pct@rev.desc.d6.uiv.p1.C.box.dual-d2.resid-agree-stable\endcsname{0.0}
\expandafter\def\csname ppt@rev.desc.d6.uiv.p1.C.box.dual-d2.resid-agree-stable\endcsname{0.0}
\expandafter\def\csname val@rev.desc.d6.uiv.p1.C.box.dual-d2.resid-disagree\endcsname{0.265}
\expandafter\def\csname cnt@rev.desc.d6.uiv.p1.C.box.dual-d2.resid-disagree\endcsname{0}
\expandafter\def\csname pct@rev.desc.d6.uiv.p1.C.box.dual-d2.resid-disagree\endcsname{26.5}
\expandafter\def\csname ppt@rev.desc.d6.uiv.p1.C.box.dual-d2.resid-disagree\endcsname{$+$26.5}
\expandafter\def\csname val@rev.desc.d6.uiv.p1.C.box.dual-d2.share-agree-stable\endcsname{0.583}
\expandafter\def\csname cnt@rev.desc.d6.uiv.p1.C.box.dual-d2.share-agree-stable\endcsname{1}
\expandafter\def\csname pct@rev.desc.d6.uiv.p1.C.box.dual-d2.share-agree-stable\endcsname{58.3}
\expandafter\def\csname ppt@rev.desc.d6.uiv.p1.C.box.dual-d2.share-agree-stable\endcsname{$+$58.3}
\expandafter\def\csname val@rev.desc.d6.uiv.p1.C.box.grid2-d3.N-agree-stable\endcsname{50}
\expandafter\def\csname cnt@rev.desc.d6.uiv.p1.C.box.grid2-d3.N-agree-stable\endcsname{50}
\expandafter\def\csname pct@rev.desc.d6.uiv.p1.C.box.grid2-d3.N-agree-stable\endcsname{5000.0}
\expandafter\def\csname ppt@rev.desc.d6.uiv.p1.C.box.grid2-d3.N-agree-stable\endcsname{$+$5000.0}
\expandafter\def\csname val@rev.desc.d6.uiv.p1.C.box.grid2-d3.N-disagree\endcsname{36}
\expandafter\def\csname cnt@rev.desc.d6.uiv.p1.C.box.grid2-d3.N-disagree\endcsname{36}
\expandafter\def\csname pct@rev.desc.d6.uiv.p1.C.box.grid2-d3.N-disagree\endcsname{3600.0}
\expandafter\def\csname ppt@rev.desc.d6.uiv.p1.C.box.grid2-d3.N-disagree\endcsname{$+$3600.0}
\expandafter\def\csname val@rev.desc.d6.uiv.p1.C.box.grid2-d3.boot-resid.ci95.0\endcsname{-0.058}
\expandafter\def\csname cnt@rev.desc.d6.uiv.p1.C.box.grid2-d3.boot-resid.ci95.0\endcsname{0}
\expandafter\def\csname pct@rev.desc.d6.uiv.p1.C.box.grid2-d3.boot-resid.ci95.0\endcsname{$-$5.8}
\expandafter\def\csname ppt@rev.desc.d6.uiv.p1.C.box.grid2-d3.boot-resid.ci95.0\endcsname{$-$5.8}
\expandafter\def\csname val@rev.desc.d6.uiv.p1.C.box.grid2-d3.boot-resid.ci95.1\endcsname{0.000}
\expandafter\def\csname cnt@rev.desc.d6.uiv.p1.C.box.grid2-d3.boot-resid.ci95.1\endcsname{0}
\expandafter\def\csname pct@rev.desc.d6.uiv.p1.C.box.grid2-d3.boot-resid.ci95.1\endcsname{0.0}
\expandafter\def\csname ppt@rev.desc.d6.uiv.p1.C.box.grid2-d3.boot-resid.ci95.1\endcsname{0.0}
\expandafter\def\csname val@rev.desc.d6.uiv.p1.C.box.grid2-d3.boot-resid.n-apps\endcsname{12}
\expandafter\def\csname cnt@rev.desc.d6.uiv.p1.C.box.grid2-d3.boot-resid.n-apps\endcsname{12}
\expandafter\def\csname pct@rev.desc.d6.uiv.p1.C.box.grid2-d3.boot-resid.n-apps\endcsname{1200.0}
\expandafter\def\csname ppt@rev.desc.d6.uiv.p1.C.box.grid2-d3.boot-resid.n-apps\endcsname{$+$1200.0}
\expandafter\def\csname val@rev.desc.d6.uiv.p1.C.box.grid2-d3.boot-resid.n-boot-valid\endcsname{4000}
\expandafter\def\csname cnt@rev.desc.d6.uiv.p1.C.box.grid2-d3.boot-resid.n-boot-valid\endcsname{4{,}000}
\expandafter\def\csname pct@rev.desc.d6.uiv.p1.C.box.grid2-d3.boot-resid.n-boot-valid\endcsname{400000.0}
\expandafter\def\csname ppt@rev.desc.d6.uiv.p1.C.box.grid2-d3.boot-resid.n-boot-valid\endcsname{$+$400000.0}
\expandafter\def\csname val@rev.desc.d6.uiv.p1.C.box.grid2-d3.boot-resid.value\endcsname{-0.020}
\expandafter\def\csname cnt@rev.desc.d6.uiv.p1.C.box.grid2-d3.boot-resid.value\endcsname{0}
\expandafter\def\csname pct@rev.desc.d6.uiv.p1.C.box.grid2-d3.boot-resid.value\endcsname{$-$2.0}
\expandafter\def\csname ppt@rev.desc.d6.uiv.p1.C.box.grid2-d3.boot-resid.value\endcsname{$-$2.0}
\expandafter\def\csname val@rev.desc.d6.uiv.p1.C.box.grid2-d3.boot-share.ci95.0\endcsname{0.448}
\expandafter\def\csname cnt@rev.desc.d6.uiv.p1.C.box.grid2-d3.boot-share.ci95.0\endcsname{0}
\expandafter\def\csname pct@rev.desc.d6.uiv.p1.C.box.grid2-d3.boot-share.ci95.0\endcsname{44.8}
\expandafter\def\csname ppt@rev.desc.d6.uiv.p1.C.box.grid2-d3.boot-share.ci95.0\endcsname{$+$44.8}
\expandafter\def\csname val@rev.desc.d6.uiv.p1.C.box.grid2-d3.boot-share.ci95.1\endcsname{0.594}
\expandafter\def\csname cnt@rev.desc.d6.uiv.p1.C.box.grid2-d3.boot-share.ci95.1\endcsname{1}
\expandafter\def\csname pct@rev.desc.d6.uiv.p1.C.box.grid2-d3.boot-share.ci95.1\endcsname{59.4}
\expandafter\def\csname ppt@rev.desc.d6.uiv.p1.C.box.grid2-d3.boot-share.ci95.1\endcsname{$+$59.4}
\expandafter\def\csname val@rev.desc.d6.uiv.p1.C.box.grid2-d3.boot-share.n-apps\endcsname{12}
\expandafter\def\csname cnt@rev.desc.d6.uiv.p1.C.box.grid2-d3.boot-share.n-apps\endcsname{12}
\expandafter\def\csname pct@rev.desc.d6.uiv.p1.C.box.grid2-d3.boot-share.n-apps\endcsname{1200.0}
\expandafter\def\csname ppt@rev.desc.d6.uiv.p1.C.box.grid2-d3.boot-share.n-apps\endcsname{$+$1200.0}
\expandafter\def\csname val@rev.desc.d6.uiv.p1.C.box.grid2-d3.boot-share.n-boot-valid\endcsname{4000}
\expandafter\def\csname cnt@rev.desc.d6.uiv.p1.C.box.grid2-d3.boot-share.n-boot-valid\endcsname{4{,}000}
\expandafter\def\csname pct@rev.desc.d6.uiv.p1.C.box.grid2-d3.boot-share.n-boot-valid\endcsname{400000.0}
\expandafter\def\csname ppt@rev.desc.d6.uiv.p1.C.box.grid2-d3.boot-share.n-boot-valid\endcsname{$+$400000.0}
\expandafter\def\csname val@rev.desc.d6.uiv.p1.C.box.grid2-d3.boot-share.value\endcsname{0.521}
\expandafter\def\csname cnt@rev.desc.d6.uiv.p1.C.box.grid2-d3.boot-share.value\endcsname{1}
\expandafter\def\csname pct@rev.desc.d6.uiv.p1.C.box.grid2-d3.boot-share.value\endcsname{52.1}
\expandafter\def\csname ppt@rev.desc.d6.uiv.p1.C.box.grid2-d3.boot-share.value\endcsname{$+$52.1}
\expandafter\def\csname val@rev.desc.d6.uiv.p1.C.box.grid2-d3.n\endcsname{96}
\expandafter\def\csname cnt@rev.desc.d6.uiv.p1.C.box.grid2-d3.n\endcsname{96}
\expandafter\def\csname pct@rev.desc.d6.uiv.p1.C.box.grid2-d3.n\endcsname{9600.0}
\expandafter\def\csname ppt@rev.desc.d6.uiv.p1.C.box.grid2-d3.n\endcsname{$+$9600.0}
\expandafter\def\csname val@rev.desc.d6.uiv.p1.C.box.grid2-d3.resid-agree-stable\endcsname{-0.020}
\expandafter\def\csname cnt@rev.desc.d6.uiv.p1.C.box.grid2-d3.resid-agree-stable\endcsname{0}
\expandafter\def\csname pct@rev.desc.d6.uiv.p1.C.box.grid2-d3.resid-agree-stable\endcsname{$-$2.0}
\expandafter\def\csname ppt@rev.desc.d6.uiv.p1.C.box.grid2-d3.resid-agree-stable\endcsname{$-$2.0}
\expandafter\def\csname val@rev.desc.d6.uiv.p1.C.box.grid2-d3.resid-disagree\endcsname{0.306}
\expandafter\def\csname cnt@rev.desc.d6.uiv.p1.C.box.grid2-d3.resid-disagree\endcsname{0}
\expandafter\def\csname pct@rev.desc.d6.uiv.p1.C.box.grid2-d3.resid-disagree\endcsname{30.6}
\expandafter\def\csname ppt@rev.desc.d6.uiv.p1.C.box.grid2-d3.resid-disagree\endcsname{$+$30.6}
\expandafter\def\csname val@rev.desc.d6.uiv.p1.C.box.grid2-d3.share-agree-stable\endcsname{0.521}
\expandafter\def\csname cnt@rev.desc.d6.uiv.p1.C.box.grid2-d3.share-agree-stable\endcsname{1}
\expandafter\def\csname pct@rev.desc.d6.uiv.p1.C.box.grid2-d3.share-agree-stable\endcsname{52.1}
\expandafter\def\csname ppt@rev.desc.d6.uiv.p1.C.box.grid2-d3.share-agree-stable\endcsname{$+$52.1}
\expandafter\def\csname val@rev.desc.d6.uiv.p1.C.box.zoomclick-d3.N-agree-stable\endcsname{61}
\expandafter\def\csname cnt@rev.desc.d6.uiv.p1.C.box.zoomclick-d3.N-agree-stable\endcsname{61}
\expandafter\def\csname pct@rev.desc.d6.uiv.p1.C.box.zoomclick-d3.N-agree-stable\endcsname{6100.0}
\expandafter\def\csname ppt@rev.desc.d6.uiv.p1.C.box.zoomclick-d3.N-agree-stable\endcsname{$+$6100.0}
\expandafter\def\csname val@rev.desc.d6.uiv.p1.C.box.zoomclick-d3.N-disagree\endcsname{28}
\expandafter\def\csname cnt@rev.desc.d6.uiv.p1.C.box.zoomclick-d3.N-disagree\endcsname{28}
\expandafter\def\csname pct@rev.desc.d6.uiv.p1.C.box.zoomclick-d3.N-disagree\endcsname{2800.0}
\expandafter\def\csname ppt@rev.desc.d6.uiv.p1.C.box.zoomclick-d3.N-disagree\endcsname{$+$2800.0}
\expandafter\def\csname val@rev.desc.d6.uiv.p1.C.box.zoomclick-d3.boot-resid.ci95.0\endcsname{-0.079}
\expandafter\def\csname cnt@rev.desc.d6.uiv.p1.C.box.zoomclick-d3.boot-resid.ci95.0\endcsname{0}
\expandafter\def\csname pct@rev.desc.d6.uiv.p1.C.box.zoomclick-d3.boot-resid.ci95.0\endcsname{$-$7.9}
\expandafter\def\csname ppt@rev.desc.d6.uiv.p1.C.box.zoomclick-d3.boot-resid.ci95.0\endcsname{$-$7.9}
\expandafter\def\csname val@rev.desc.d6.uiv.p1.C.box.zoomclick-d3.boot-resid.ci95.1\endcsname{0.000}
\expandafter\def\csname cnt@rev.desc.d6.uiv.p1.C.box.zoomclick-d3.boot-resid.ci95.1\endcsname{0}
\expandafter\def\csname pct@rev.desc.d6.uiv.p1.C.box.zoomclick-d3.boot-resid.ci95.1\endcsname{0.0}
\expandafter\def\csname ppt@rev.desc.d6.uiv.p1.C.box.zoomclick-d3.boot-resid.ci95.1\endcsname{0.0}
\expandafter\def\csname val@rev.desc.d6.uiv.p1.C.box.zoomclick-d3.boot-resid.n-apps\endcsname{12}
\expandafter\def\csname cnt@rev.desc.d6.uiv.p1.C.box.zoomclick-d3.boot-resid.n-apps\endcsname{12}
\expandafter\def\csname pct@rev.desc.d6.uiv.p1.C.box.zoomclick-d3.boot-resid.n-apps\endcsname{1200.0}
\expandafter\def\csname ppt@rev.desc.d6.uiv.p1.C.box.zoomclick-d3.boot-resid.n-apps\endcsname{$+$1200.0}
\expandafter\def\csname val@rev.desc.d6.uiv.p1.C.box.zoomclick-d3.boot-resid.n-boot-valid\endcsname{4000}
\expandafter\def\csname cnt@rev.desc.d6.uiv.p1.C.box.zoomclick-d3.boot-resid.n-boot-valid\endcsname{4{,}000}
\expandafter\def\csname pct@rev.desc.d6.uiv.p1.C.box.zoomclick-d3.boot-resid.n-boot-valid\endcsname{400000.0}
\expandafter\def\csname ppt@rev.desc.d6.uiv.p1.C.box.zoomclick-d3.boot-resid.n-boot-valid\endcsname{$+$400000.0}
\expandafter\def\csname val@rev.desc.d6.uiv.p1.C.box.zoomclick-d3.boot-resid.value\endcsname{-0.033}
\expandafter\def\csname cnt@rev.desc.d6.uiv.p1.C.box.zoomclick-d3.boot-resid.value\endcsname{0}
\expandafter\def\csname pct@rev.desc.d6.uiv.p1.C.box.zoomclick-d3.boot-resid.value\endcsname{$-$3.3}
\expandafter\def\csname ppt@rev.desc.d6.uiv.p1.C.box.zoomclick-d3.boot-resid.value\endcsname{$-$3.3}
\expandafter\def\csname val@rev.desc.d6.uiv.p1.C.box.zoomclick-d3.boot-share.ci95.0\endcsname{0.562}
\expandafter\def\csname cnt@rev.desc.d6.uiv.p1.C.box.zoomclick-d3.boot-share.ci95.0\endcsname{1}
\expandafter\def\csname pct@rev.desc.d6.uiv.p1.C.box.zoomclick-d3.boot-share.ci95.0\endcsname{56.2}
\expandafter\def\csname ppt@rev.desc.d6.uiv.p1.C.box.zoomclick-d3.boot-share.ci95.0\endcsname{$+$56.2}
\expandafter\def\csname val@rev.desc.d6.uiv.p1.C.box.zoomclick-d3.boot-share.ci95.1\endcsname{0.708}
\expandafter\def\csname cnt@rev.desc.d6.uiv.p1.C.box.zoomclick-d3.boot-share.ci95.1\endcsname{1}
\expandafter\def\csname pct@rev.desc.d6.uiv.p1.C.box.zoomclick-d3.boot-share.ci95.1\endcsname{70.8}
\expandafter\def\csname ppt@rev.desc.d6.uiv.p1.C.box.zoomclick-d3.boot-share.ci95.1\endcsname{$+$70.8}
\expandafter\def\csname val@rev.desc.d6.uiv.p1.C.box.zoomclick-d3.boot-share.n-apps\endcsname{12}
\expandafter\def\csname cnt@rev.desc.d6.uiv.p1.C.box.zoomclick-d3.boot-share.n-apps\endcsname{12}
\expandafter\def\csname pct@rev.desc.d6.uiv.p1.C.box.zoomclick-d3.boot-share.n-apps\endcsname{1200.0}
\expandafter\def\csname ppt@rev.desc.d6.uiv.p1.C.box.zoomclick-d3.boot-share.n-apps\endcsname{$+$1200.0}
\expandafter\def\csname val@rev.desc.d6.uiv.p1.C.box.zoomclick-d3.boot-share.n-boot-valid\endcsname{4000}
\expandafter\def\csname cnt@rev.desc.d6.uiv.p1.C.box.zoomclick-d3.boot-share.n-boot-valid\endcsname{4{,}000}
\expandafter\def\csname pct@rev.desc.d6.uiv.p1.C.box.zoomclick-d3.boot-share.n-boot-valid\endcsname{400000.0}
\expandafter\def\csname ppt@rev.desc.d6.uiv.p1.C.box.zoomclick-d3.boot-share.n-boot-valid\endcsname{$+$400000.0}
\expandafter\def\csname val@rev.desc.d6.uiv.p1.C.box.zoomclick-d3.boot-share.value\endcsname{0.635}
\expandafter\def\csname cnt@rev.desc.d6.uiv.p1.C.box.zoomclick-d3.boot-share.value\endcsname{1}
\expandafter\def\csname pct@rev.desc.d6.uiv.p1.C.box.zoomclick-d3.boot-share.value\endcsname{63.5}
\expandafter\def\csname ppt@rev.desc.d6.uiv.p1.C.box.zoomclick-d3.boot-share.value\endcsname{$+$63.5}
\expandafter\def\csname val@rev.desc.d6.uiv.p1.C.box.zoomclick-d3.n\endcsname{96}
\expandafter\def\csname cnt@rev.desc.d6.uiv.p1.C.box.zoomclick-d3.n\endcsname{96}
\expandafter\def\csname pct@rev.desc.d6.uiv.p1.C.box.zoomclick-d3.n\endcsname{9600.0}
\expandafter\def\csname ppt@rev.desc.d6.uiv.p1.C.box.zoomclick-d3.n\endcsname{$+$9600.0}
\expandafter\def\csname val@rev.desc.d6.uiv.p1.C.box.zoomclick-d3.resid-agree-stable\endcsname{-0.033}
\expandafter\def\csname cnt@rev.desc.d6.uiv.p1.C.box.zoomclick-d3.resid-agree-stable\endcsname{0}
\expandafter\def\csname pct@rev.desc.d6.uiv.p1.C.box.zoomclick-d3.resid-agree-stable\endcsname{$-$3.3}
\expandafter\def\csname ppt@rev.desc.d6.uiv.p1.C.box.zoomclick-d3.resid-agree-stable\endcsname{$-$3.3}
\expandafter\def\csname val@rev.desc.d6.uiv.p1.C.box.zoomclick-d3.resid-disagree\endcsname{0.107}
\expandafter\def\csname cnt@rev.desc.d6.uiv.p1.C.box.zoomclick-d3.resid-disagree\endcsname{0}
\expandafter\def\csname pct@rev.desc.d6.uiv.p1.C.box.zoomclick-d3.resid-disagree\endcsname{10.7}
\expandafter\def\csname ppt@rev.desc.d6.uiv.p1.C.box.zoomclick-d3.resid-disagree\endcsname{$+$10.7}
\expandafter\def\csname val@rev.desc.d6.uiv.p1.C.box.zoomclick-d3.share-agree-stable\endcsname{0.635}
\expandafter\def\csname cnt@rev.desc.d6.uiv.p1.C.box.zoomclick-d3.share-agree-stable\endcsname{1}
\expandafter\def\csname pct@rev.desc.d6.uiv.p1.C.box.zoomclick-d3.share-agree-stable\endcsname{63.5}
\expandafter\def\csname ppt@rev.desc.d6.uiv.p1.C.box.zoomclick-d3.share-agree-stable\endcsname{$+$63.5}
\expandafter\def\csname val@rev.desc.d6.uiv.p1.C.tdiag.catalog-pooled.N-agree-stable\endcsname{276}
\expandafter\def\csname cnt@rev.desc.d6.uiv.p1.C.tdiag.catalog-pooled.N-agree-stable\endcsname{276}
\expandafter\def\csname pct@rev.desc.d6.uiv.p1.C.tdiag.catalog-pooled.N-agree-stable\endcsname{27600.0}
\expandafter\def\csname ppt@rev.desc.d6.uiv.p1.C.tdiag.catalog-pooled.N-agree-stable\endcsname{$+$27600.0}
\expandafter\def\csname val@rev.desc.d6.uiv.p1.C.tdiag.catalog-pooled.N-disagree\endcsname{65}
\expandafter\def\csname cnt@rev.desc.d6.uiv.p1.C.tdiag.catalog-pooled.N-disagree\endcsname{65}
\expandafter\def\csname pct@rev.desc.d6.uiv.p1.C.tdiag.catalog-pooled.N-disagree\endcsname{6500.0}
\expandafter\def\csname ppt@rev.desc.d6.uiv.p1.C.tdiag.catalog-pooled.N-disagree\endcsname{$+$6500.0}
\expandafter\def\csname val@rev.desc.d6.uiv.p1.C.tdiag.catalog-pooled.boot-resid.ci95.0\endcsname{0.000}
\expandafter\def\csname cnt@rev.desc.d6.uiv.p1.C.tdiag.catalog-pooled.boot-resid.ci95.0\endcsname{0}
\expandafter\def\csname pct@rev.desc.d6.uiv.p1.C.tdiag.catalog-pooled.boot-resid.ci95.0\endcsname{0.0}
\expandafter\def\csname ppt@rev.desc.d6.uiv.p1.C.tdiag.catalog-pooled.boot-resid.ci95.0\endcsname{0.0}
\expandafter\def\csname val@rev.desc.d6.uiv.p1.C.tdiag.catalog-pooled.boot-resid.ci95.1\endcsname{0.054}
\expandafter\def\csname cnt@rev.desc.d6.uiv.p1.C.tdiag.catalog-pooled.boot-resid.ci95.1\endcsname{0}
\expandafter\def\csname pct@rev.desc.d6.uiv.p1.C.tdiag.catalog-pooled.boot-resid.ci95.1\endcsname{5.4}
\expandafter\def\csname ppt@rev.desc.d6.uiv.p1.C.tdiag.catalog-pooled.boot-resid.ci95.1\endcsname{$+$5.4}
\expandafter\def\csname val@rev.desc.d6.uiv.p1.C.tdiag.catalog-pooled.boot-resid.n-apps\endcsname{12}
\expandafter\def\csname cnt@rev.desc.d6.uiv.p1.C.tdiag.catalog-pooled.boot-resid.n-apps\endcsname{12}
\expandafter\def\csname pct@rev.desc.d6.uiv.p1.C.tdiag.catalog-pooled.boot-resid.n-apps\endcsname{1200.0}
\expandafter\def\csname ppt@rev.desc.d6.uiv.p1.C.tdiag.catalog-pooled.boot-resid.n-apps\endcsname{$+$1200.0}
\expandafter\def\csname val@rev.desc.d6.uiv.p1.C.tdiag.catalog-pooled.boot-resid.n-boot-valid\endcsname{4000}
\expandafter\def\csname cnt@rev.desc.d6.uiv.p1.C.tdiag.catalog-pooled.boot-resid.n-boot-valid\endcsname{4{,}000}
\expandafter\def\csname pct@rev.desc.d6.uiv.p1.C.tdiag.catalog-pooled.boot-resid.n-boot-valid\endcsname{400000.0}
\expandafter\def\csname ppt@rev.desc.d6.uiv.p1.C.tdiag.catalog-pooled.boot-resid.n-boot-valid\endcsname{$+$400000.0}
\expandafter\def\csname val@rev.desc.d6.uiv.p1.C.tdiag.catalog-pooled.boot-resid.value\endcsname{0.025}
\expandafter\def\csname cnt@rev.desc.d6.uiv.p1.C.tdiag.catalog-pooled.boot-resid.value\endcsname{0}
\expandafter\def\csname pct@rev.desc.d6.uiv.p1.C.tdiag.catalog-pooled.boot-resid.value\endcsname{2.5}
\expandafter\def\csname ppt@rev.desc.d6.uiv.p1.C.tdiag.catalog-pooled.boot-resid.value\endcsname{$+$2.5}
\expandafter\def\csname val@rev.desc.d6.uiv.p1.C.tdiag.catalog-pooled.boot-share.ci95.0\endcsname{0.659}
\expandafter\def\csname cnt@rev.desc.d6.uiv.p1.C.tdiag.catalog-pooled.boot-share.ci95.0\endcsname{1}
\expandafter\def\csname pct@rev.desc.d6.uiv.p1.C.tdiag.catalog-pooled.boot-share.ci95.0\endcsname{65.9}
\expandafter\def\csname ppt@rev.desc.d6.uiv.p1.C.tdiag.catalog-pooled.boot-share.ci95.0\endcsname{$+$65.9}
\expandafter\def\csname val@rev.desc.d6.uiv.p1.C.tdiag.catalog-pooled.boot-share.ci95.1\endcsname{0.781}
\expandafter\def\csname cnt@rev.desc.d6.uiv.p1.C.tdiag.catalog-pooled.boot-share.ci95.1\endcsname{1}
\expandafter\def\csname pct@rev.desc.d6.uiv.p1.C.tdiag.catalog-pooled.boot-share.ci95.1\endcsname{78.1}
\expandafter\def\csname ppt@rev.desc.d6.uiv.p1.C.tdiag.catalog-pooled.boot-share.ci95.1\endcsname{$+$78.1}
\expandafter\def\csname val@rev.desc.d6.uiv.p1.C.tdiag.catalog-pooled.boot-share.n-apps\endcsname{12}
\expandafter\def\csname cnt@rev.desc.d6.uiv.p1.C.tdiag.catalog-pooled.boot-share.n-apps\endcsname{12}
\expandafter\def\csname pct@rev.desc.d6.uiv.p1.C.tdiag.catalog-pooled.boot-share.n-apps\endcsname{1200.0}
\expandafter\def\csname ppt@rev.desc.d6.uiv.p1.C.tdiag.catalog-pooled.boot-share.n-apps\endcsname{$+$1200.0}
\expandafter\def\csname val@rev.desc.d6.uiv.p1.C.tdiag.catalog-pooled.boot-share.n-boot-valid\endcsname{4000}
\expandafter\def\csname cnt@rev.desc.d6.uiv.p1.C.tdiag.catalog-pooled.boot-share.n-boot-valid\endcsname{4{,}000}
\expandafter\def\csname pct@rev.desc.d6.uiv.p1.C.tdiag.catalog-pooled.boot-share.n-boot-valid\endcsname{400000.0}
\expandafter\def\csname ppt@rev.desc.d6.uiv.p1.C.tdiag.catalog-pooled.boot-share.n-boot-valid\endcsname{$+$400000.0}
\expandafter\def\csname val@rev.desc.d6.uiv.p1.C.tdiag.catalog-pooled.boot-share.value\endcsname{0.719}
\expandafter\def\csname cnt@rev.desc.d6.uiv.p1.C.tdiag.catalog-pooled.boot-share.value\endcsname{1}
\expandafter\def\csname pct@rev.desc.d6.uiv.p1.C.tdiag.catalog-pooled.boot-share.value\endcsname{71.9}
\expandafter\def\csname ppt@rev.desc.d6.uiv.p1.C.tdiag.catalog-pooled.boot-share.value\endcsname{$+$71.9}
\expandafter\def\csname val@rev.desc.d6.uiv.p1.C.tdiag.catalog-pooled.gap\endcsname{0.236}
\expandafter\def\csname cnt@rev.desc.d6.uiv.p1.C.tdiag.catalog-pooled.gap\endcsname{0}
\expandafter\def\csname pct@rev.desc.d6.uiv.p1.C.tdiag.catalog-pooled.gap\endcsname{23.6}
\expandafter\def\csname ppt@rev.desc.d6.uiv.p1.C.tdiag.catalog-pooled.gap\endcsname{$+$23.6}
\expandafter\def\csname val@rev.desc.d6.uiv.p1.C.tdiag.catalog-pooled.n\endcsname{384}
\expandafter\def\csname cnt@rev.desc.d6.uiv.p1.C.tdiag.catalog-pooled.n\endcsname{384}
\expandafter\def\csname pct@rev.desc.d6.uiv.p1.C.tdiag.catalog-pooled.n\endcsname{38400.0}
\expandafter\def\csname ppt@rev.desc.d6.uiv.p1.C.tdiag.catalog-pooled.n\endcsname{$+$38400.0}
\expandafter\def\csname val@rev.desc.d6.uiv.p1.C.tdiag.catalog-pooled.resid-agree-stable\endcsname{0.025}
\expandafter\def\csname cnt@rev.desc.d6.uiv.p1.C.tdiag.catalog-pooled.resid-agree-stable\endcsname{0}
\expandafter\def\csname pct@rev.desc.d6.uiv.p1.C.tdiag.catalog-pooled.resid-agree-stable\endcsname{2.5}
\expandafter\def\csname ppt@rev.desc.d6.uiv.p1.C.tdiag.catalog-pooled.resid-agree-stable\endcsname{$+$2.5}
\expandafter\def\csname val@rev.desc.d6.uiv.p1.C.tdiag.catalog-pooled.resid-disagree\endcsname{0.262}
\expandafter\def\csname cnt@rev.desc.d6.uiv.p1.C.tdiag.catalog-pooled.resid-disagree\endcsname{0}
\expandafter\def\csname pct@rev.desc.d6.uiv.p1.C.tdiag.catalog-pooled.resid-disagree\endcsname{26.2}
\expandafter\def\csname ppt@rev.desc.d6.uiv.p1.C.tdiag.catalog-pooled.resid-disagree\endcsname{$+$26.2}
\expandafter\def\csname val@rev.desc.d6.uiv.p1.C.tdiag.catalog-pooled.share-agree-stable\endcsname{0.719}
\expandafter\def\csname cnt@rev.desc.d6.uiv.p1.C.tdiag.catalog-pooled.share-agree-stable\endcsname{1}
\expandafter\def\csname pct@rev.desc.d6.uiv.p1.C.tdiag.catalog-pooled.share-agree-stable\endcsname{71.9}
\expandafter\def\csname ppt@rev.desc.d6.uiv.p1.C.tdiag.catalog-pooled.share-agree-stable\endcsname{$+$71.9}
\expandafter\def\csname val@rev.desc.d6.uiv.p1.C.tdiag.center-d1.N-agree-stable\endcsname{71}
\expandafter\def\csname cnt@rev.desc.d6.uiv.p1.C.tdiag.center-d1.N-agree-stable\endcsname{71}
\expandafter\def\csname pct@rev.desc.d6.uiv.p1.C.tdiag.center-d1.N-agree-stable\endcsname{7100.0}
\expandafter\def\csname ppt@rev.desc.d6.uiv.p1.C.tdiag.center-d1.N-agree-stable\endcsname{$+$7100.0}
\expandafter\def\csname val@rev.desc.d6.uiv.p1.C.tdiag.center-d1.N-disagree\endcsname{15}
\expandafter\def\csname cnt@rev.desc.d6.uiv.p1.C.tdiag.center-d1.N-disagree\endcsname{15}
\expandafter\def\csname pct@rev.desc.d6.uiv.p1.C.tdiag.center-d1.N-disagree\endcsname{1500.0}
\expandafter\def\csname ppt@rev.desc.d6.uiv.p1.C.tdiag.center-d1.N-disagree\endcsname{$+$1500.0}
\expandafter\def\csname val@rev.desc.d6.uiv.p1.C.tdiag.center-d1.boot-resid.ci95.0\endcsname{0.000}
\expandafter\def\csname cnt@rev.desc.d6.uiv.p1.C.tdiag.center-d1.boot-resid.ci95.0\endcsname{0}
\expandafter\def\csname pct@rev.desc.d6.uiv.p1.C.tdiag.center-d1.boot-resid.ci95.0\endcsname{0.0}
\expandafter\def\csname ppt@rev.desc.d6.uiv.p1.C.tdiag.center-d1.boot-resid.ci95.0\endcsname{0.0}
\expandafter\def\csname val@rev.desc.d6.uiv.p1.C.tdiag.center-d1.boot-resid.ci95.1\endcsname{0.065}
\expandafter\def\csname cnt@rev.desc.d6.uiv.p1.C.tdiag.center-d1.boot-resid.ci95.1\endcsname{0}
\expandafter\def\csname pct@rev.desc.d6.uiv.p1.C.tdiag.center-d1.boot-resid.ci95.1\endcsname{6.5}
\expandafter\def\csname ppt@rev.desc.d6.uiv.p1.C.tdiag.center-d1.boot-resid.ci95.1\endcsname{$+$6.5}
\expandafter\def\csname val@rev.desc.d6.uiv.p1.C.tdiag.center-d1.boot-resid.n-apps\endcsname{12}
\expandafter\def\csname cnt@rev.desc.d6.uiv.p1.C.tdiag.center-d1.boot-resid.n-apps\endcsname{12}
\expandafter\def\csname pct@rev.desc.d6.uiv.p1.C.tdiag.center-d1.boot-resid.n-apps\endcsname{1200.0}
\expandafter\def\csname ppt@rev.desc.d6.uiv.p1.C.tdiag.center-d1.boot-resid.n-apps\endcsname{$+$1200.0}
\expandafter\def\csname val@rev.desc.d6.uiv.p1.C.tdiag.center-d1.boot-resid.n-boot-valid\endcsname{4000}
\expandafter\def\csname cnt@rev.desc.d6.uiv.p1.C.tdiag.center-d1.boot-resid.n-boot-valid\endcsname{4{,}000}
\expandafter\def\csname pct@rev.desc.d6.uiv.p1.C.tdiag.center-d1.boot-resid.n-boot-valid\endcsname{400000.0}
\expandafter\def\csname ppt@rev.desc.d6.uiv.p1.C.tdiag.center-d1.boot-resid.n-boot-valid\endcsname{$+$400000.0}
\expandafter\def\csname val@rev.desc.d6.uiv.p1.C.tdiag.center-d1.boot-resid.value\endcsname{0.028}
\expandafter\def\csname cnt@rev.desc.d6.uiv.p1.C.tdiag.center-d1.boot-resid.value\endcsname{0}
\expandafter\def\csname pct@rev.desc.d6.uiv.p1.C.tdiag.center-d1.boot-resid.value\endcsname{2.8}
\expandafter\def\csname ppt@rev.desc.d6.uiv.p1.C.tdiag.center-d1.boot-resid.value\endcsname{$+$2.8}
\expandafter\def\csname val@rev.desc.d6.uiv.p1.C.tdiag.center-d1.boot-share.ci95.0\endcsname{0.667}
\expandafter\def\csname cnt@rev.desc.d6.uiv.p1.C.tdiag.center-d1.boot-share.ci95.0\endcsname{1}
\expandafter\def\csname pct@rev.desc.d6.uiv.p1.C.tdiag.center-d1.boot-share.ci95.0\endcsname{66.7}
\expandafter\def\csname ppt@rev.desc.d6.uiv.p1.C.tdiag.center-d1.boot-share.ci95.0\endcsname{$+$66.7}
\expandafter\def\csname val@rev.desc.d6.uiv.p1.C.tdiag.center-d1.boot-share.ci95.1\endcsname{0.812}
\expandafter\def\csname cnt@rev.desc.d6.uiv.p1.C.tdiag.center-d1.boot-share.ci95.1\endcsname{1}
\expandafter\def\csname pct@rev.desc.d6.uiv.p1.C.tdiag.center-d1.boot-share.ci95.1\endcsname{81.2}
\expandafter\def\csname ppt@rev.desc.d6.uiv.p1.C.tdiag.center-d1.boot-share.ci95.1\endcsname{$+$81.2}
\expandafter\def\csname val@rev.desc.d6.uiv.p1.C.tdiag.center-d1.boot-share.n-apps\endcsname{12}
\expandafter\def\csname cnt@rev.desc.d6.uiv.p1.C.tdiag.center-d1.boot-share.n-apps\endcsname{12}
\expandafter\def\csname pct@rev.desc.d6.uiv.p1.C.tdiag.center-d1.boot-share.n-apps\endcsname{1200.0}
\expandafter\def\csname ppt@rev.desc.d6.uiv.p1.C.tdiag.center-d1.boot-share.n-apps\endcsname{$+$1200.0}
\expandafter\def\csname val@rev.desc.d6.uiv.p1.C.tdiag.center-d1.boot-share.n-boot-valid\endcsname{4000}
\expandafter\def\csname cnt@rev.desc.d6.uiv.p1.C.tdiag.center-d1.boot-share.n-boot-valid\endcsname{4{,}000}
\expandafter\def\csname pct@rev.desc.d6.uiv.p1.C.tdiag.center-d1.boot-share.n-boot-valid\endcsname{400000.0}
\expandafter\def\csname ppt@rev.desc.d6.uiv.p1.C.tdiag.center-d1.boot-share.n-boot-valid\endcsname{$+$400000.0}
\expandafter\def\csname val@rev.desc.d6.uiv.p1.C.tdiag.center-d1.boot-share.value\endcsname{0.740}
\expandafter\def\csname cnt@rev.desc.d6.uiv.p1.C.tdiag.center-d1.boot-share.value\endcsname{1}
\expandafter\def\csname pct@rev.desc.d6.uiv.p1.C.tdiag.center-d1.boot-share.value\endcsname{74.0}
\expandafter\def\csname ppt@rev.desc.d6.uiv.p1.C.tdiag.center-d1.boot-share.value\endcsname{$+$74.0}
\expandafter\def\csname val@rev.desc.d6.uiv.p1.C.tdiag.center-d1.n\endcsname{96}
\expandafter\def\csname cnt@rev.desc.d6.uiv.p1.C.tdiag.center-d1.n\endcsname{96}
\expandafter\def\csname pct@rev.desc.d6.uiv.p1.C.tdiag.center-d1.n\endcsname{9600.0}
\expandafter\def\csname ppt@rev.desc.d6.uiv.p1.C.tdiag.center-d1.n\endcsname{$+$9600.0}
\expandafter\def\csname val@rev.desc.d6.uiv.p1.C.tdiag.center-d1.resid-agree-stable\endcsname{0.028}
\expandafter\def\csname cnt@rev.desc.d6.uiv.p1.C.tdiag.center-d1.resid-agree-stable\endcsname{0}
\expandafter\def\csname pct@rev.desc.d6.uiv.p1.C.tdiag.center-d1.resid-agree-stable\endcsname{2.8}
\expandafter\def\csname ppt@rev.desc.d6.uiv.p1.C.tdiag.center-d1.resid-agree-stable\endcsname{$+$2.8}
\expandafter\def\csname val@rev.desc.d6.uiv.p1.C.tdiag.center-d1.resid-disagree\endcsname{0.267}
\expandafter\def\csname cnt@rev.desc.d6.uiv.p1.C.tdiag.center-d1.resid-disagree\endcsname{0}
\expandafter\def\csname pct@rev.desc.d6.uiv.p1.C.tdiag.center-d1.resid-disagree\endcsname{26.7}
\expandafter\def\csname ppt@rev.desc.d6.uiv.p1.C.tdiag.center-d1.resid-disagree\endcsname{$+$26.7}
\expandafter\def\csname val@rev.desc.d6.uiv.p1.C.tdiag.center-d1.share-agree-stable\endcsname{0.740}
\expandafter\def\csname cnt@rev.desc.d6.uiv.p1.C.tdiag.center-d1.share-agree-stable\endcsname{1}
\expandafter\def\csname pct@rev.desc.d6.uiv.p1.C.tdiag.center-d1.share-agree-stable\endcsname{74.0}
\expandafter\def\csname ppt@rev.desc.d6.uiv.p1.C.tdiag.center-d1.share-agree-stable\endcsname{$+$74.0}
\expandafter\def\csname val@rev.desc.d6.uiv.p1.C.tdiag.context-d1.N-agree-stable\endcsname{73}
\expandafter\def\csname cnt@rev.desc.d6.uiv.p1.C.tdiag.context-d1.N-agree-stable\endcsname{73}
\expandafter\def\csname pct@rev.desc.d6.uiv.p1.C.tdiag.context-d1.N-agree-stable\endcsname{7300.0}
\expandafter\def\csname ppt@rev.desc.d6.uiv.p1.C.tdiag.context-d1.N-agree-stable\endcsname{$+$7300.0}
\expandafter\def\csname val@rev.desc.d6.uiv.p1.C.tdiag.context-d1.N-disagree\endcsname{12}
\expandafter\def\csname cnt@rev.desc.d6.uiv.p1.C.tdiag.context-d1.N-disagree\endcsname{12}
\expandafter\def\csname pct@rev.desc.d6.uiv.p1.C.tdiag.context-d1.N-disagree\endcsname{1200.0}
\expandafter\def\csname ppt@rev.desc.d6.uiv.p1.C.tdiag.context-d1.N-disagree\endcsname{$+$1200.0}
\expandafter\def\csname val@rev.desc.d6.uiv.p1.C.tdiag.context-d1.boot-resid.ci95.0\endcsname{0.000}
\expandafter\def\csname cnt@rev.desc.d6.uiv.p1.C.tdiag.context-d1.boot-resid.ci95.0\endcsname{0}
\expandafter\def\csname pct@rev.desc.d6.uiv.p1.C.tdiag.context-d1.boot-resid.ci95.0\endcsname{0.0}
\expandafter\def\csname ppt@rev.desc.d6.uiv.p1.C.tdiag.context-d1.boot-resid.ci95.0\endcsname{0.0}
\expandafter\def\csname val@rev.desc.d6.uiv.p1.C.tdiag.context-d1.boot-resid.ci95.1\endcsname{0.041}
\expandafter\def\csname cnt@rev.desc.d6.uiv.p1.C.tdiag.context-d1.boot-resid.ci95.1\endcsname{0}
\expandafter\def\csname pct@rev.desc.d6.uiv.p1.C.tdiag.context-d1.boot-resid.ci95.1\endcsname{4.1}
\expandafter\def\csname ppt@rev.desc.d6.uiv.p1.C.tdiag.context-d1.boot-resid.ci95.1\endcsname{$+$4.1}
\expandafter\def\csname val@rev.desc.d6.uiv.p1.C.tdiag.context-d1.boot-resid.n-apps\endcsname{12}
\expandafter\def\csname cnt@rev.desc.d6.uiv.p1.C.tdiag.context-d1.boot-resid.n-apps\endcsname{12}
\expandafter\def\csname pct@rev.desc.d6.uiv.p1.C.tdiag.context-d1.boot-resid.n-apps\endcsname{1200.0}
\expandafter\def\csname ppt@rev.desc.d6.uiv.p1.C.tdiag.context-d1.boot-resid.n-apps\endcsname{$+$1200.0}
\expandafter\def\csname val@rev.desc.d6.uiv.p1.C.tdiag.context-d1.boot-resid.n-boot-valid\endcsname{4000}
\expandafter\def\csname cnt@rev.desc.d6.uiv.p1.C.tdiag.context-d1.boot-resid.n-boot-valid\endcsname{4{,}000}
\expandafter\def\csname pct@rev.desc.d6.uiv.p1.C.tdiag.context-d1.boot-resid.n-boot-valid\endcsname{400000.0}
\expandafter\def\csname ppt@rev.desc.d6.uiv.p1.C.tdiag.context-d1.boot-resid.n-boot-valid\endcsname{$+$400000.0}
\expandafter\def\csname val@rev.desc.d6.uiv.p1.C.tdiag.context-d1.boot-resid.value\endcsname{0.014}
\expandafter\def\csname cnt@rev.desc.d6.uiv.p1.C.tdiag.context-d1.boot-resid.value\endcsname{0}
\expandafter\def\csname pct@rev.desc.d6.uiv.p1.C.tdiag.context-d1.boot-resid.value\endcsname{1.4}
\expandafter\def\csname ppt@rev.desc.d6.uiv.p1.C.tdiag.context-d1.boot-resid.value\endcsname{$+$1.4}
\expandafter\def\csname val@rev.desc.d6.uiv.p1.C.tdiag.context-d1.boot-share.ci95.0\endcsname{0.677}
\expandafter\def\csname cnt@rev.desc.d6.uiv.p1.C.tdiag.context-d1.boot-share.ci95.0\endcsname{1}
\expandafter\def\csname pct@rev.desc.d6.uiv.p1.C.tdiag.context-d1.boot-share.ci95.0\endcsname{67.7}
\expandafter\def\csname ppt@rev.desc.d6.uiv.p1.C.tdiag.context-d1.boot-share.ci95.0\endcsname{$+$67.7}
\expandafter\def\csname val@rev.desc.d6.uiv.p1.C.tdiag.context-d1.boot-share.ci95.1\endcsname{0.844}
\expandafter\def\csname cnt@rev.desc.d6.uiv.p1.C.tdiag.context-d1.boot-share.ci95.1\endcsname{1}
\expandafter\def\csname pct@rev.desc.d6.uiv.p1.C.tdiag.context-d1.boot-share.ci95.1\endcsname{84.4}
\expandafter\def\csname ppt@rev.desc.d6.uiv.p1.C.tdiag.context-d1.boot-share.ci95.1\endcsname{$+$84.4}
\expandafter\def\csname val@rev.desc.d6.uiv.p1.C.tdiag.context-d1.boot-share.n-apps\endcsname{12}
\expandafter\def\csname cnt@rev.desc.d6.uiv.p1.C.tdiag.context-d1.boot-share.n-apps\endcsname{12}
\expandafter\def\csname pct@rev.desc.d6.uiv.p1.C.tdiag.context-d1.boot-share.n-apps\endcsname{1200.0}
\expandafter\def\csname ppt@rev.desc.d6.uiv.p1.C.tdiag.context-d1.boot-share.n-apps\endcsname{$+$1200.0}
\expandafter\def\csname val@rev.desc.d6.uiv.p1.C.tdiag.context-d1.boot-share.n-boot-valid\endcsname{4000}
\expandafter\def\csname cnt@rev.desc.d6.uiv.p1.C.tdiag.context-d1.boot-share.n-boot-valid\endcsname{4{,}000}
\expandafter\def\csname pct@rev.desc.d6.uiv.p1.C.tdiag.context-d1.boot-share.n-boot-valid\endcsname{400000.0}
\expandafter\def\csname ppt@rev.desc.d6.uiv.p1.C.tdiag.context-d1.boot-share.n-boot-valid\endcsname{$+$400000.0}
\expandafter\def\csname val@rev.desc.d6.uiv.p1.C.tdiag.context-d1.boot-share.value\endcsname{0.760}
\expandafter\def\csname cnt@rev.desc.d6.uiv.p1.C.tdiag.context-d1.boot-share.value\endcsname{1}
\expandafter\def\csname pct@rev.desc.d6.uiv.p1.C.tdiag.context-d1.boot-share.value\endcsname{76.0}
\expandafter\def\csname ppt@rev.desc.d6.uiv.p1.C.tdiag.context-d1.boot-share.value\endcsname{$+$76.0}
\expandafter\def\csname val@rev.desc.d6.uiv.p1.C.tdiag.context-d1.n\endcsname{96}
\expandafter\def\csname cnt@rev.desc.d6.uiv.p1.C.tdiag.context-d1.n\endcsname{96}
\expandafter\def\csname pct@rev.desc.d6.uiv.p1.C.tdiag.context-d1.n\endcsname{9600.0}
\expandafter\def\csname ppt@rev.desc.d6.uiv.p1.C.tdiag.context-d1.n\endcsname{$+$9600.0}
\expandafter\def\csname val@rev.desc.d6.uiv.p1.C.tdiag.context-d1.resid-agree-stable\endcsname{0.014}
\expandafter\def\csname cnt@rev.desc.d6.uiv.p1.C.tdiag.context-d1.resid-agree-stable\endcsname{0}
\expandafter\def\csname pct@rev.desc.d6.uiv.p1.C.tdiag.context-d1.resid-agree-stable\endcsname{1.4}
\expandafter\def\csname ppt@rev.desc.d6.uiv.p1.C.tdiag.context-d1.resid-agree-stable\endcsname{$+$1.4}
\expandafter\def\csname val@rev.desc.d6.uiv.p1.C.tdiag.context-d1.resid-disagree\endcsname{0.000}
\expandafter\def\csname cnt@rev.desc.d6.uiv.p1.C.tdiag.context-d1.resid-disagree\endcsname{0}
\expandafter\def\csname pct@rev.desc.d6.uiv.p1.C.tdiag.context-d1.resid-disagree\endcsname{0.0}
\expandafter\def\csname ppt@rev.desc.d6.uiv.p1.C.tdiag.context-d1.resid-disagree\endcsname{0.0}
\expandafter\def\csname val@rev.desc.d6.uiv.p1.C.tdiag.context-d1.share-agree-stable\endcsname{0.760}
\expandafter\def\csname cnt@rev.desc.d6.uiv.p1.C.tdiag.context-d1.share-agree-stable\endcsname{1}
\expandafter\def\csname pct@rev.desc.d6.uiv.p1.C.tdiag.context-d1.share-agree-stable\endcsname{76.0}
\expandafter\def\csname ppt@rev.desc.d6.uiv.p1.C.tdiag.context-d1.share-agree-stable\endcsname{$+$76.0}
\expandafter\def\csname val@rev.desc.d6.uiv.p1.C.tdiag.dual-d2.N-agree-stable\endcsname{67}
\expandafter\def\csname cnt@rev.desc.d6.uiv.p1.C.tdiag.dual-d2.N-agree-stable\endcsname{67}
\expandafter\def\csname pct@rev.desc.d6.uiv.p1.C.tdiag.dual-d2.N-agree-stable\endcsname{6700.0}
\expandafter\def\csname ppt@rev.desc.d6.uiv.p1.C.tdiag.dual-d2.N-agree-stable\endcsname{$+$6700.0}
\expandafter\def\csname val@rev.desc.d6.uiv.p1.C.tdiag.dual-d2.N-disagree\endcsname{19}
\expandafter\def\csname cnt@rev.desc.d6.uiv.p1.C.tdiag.dual-d2.N-disagree\endcsname{19}
\expandafter\def\csname pct@rev.desc.d6.uiv.p1.C.tdiag.dual-d2.N-disagree\endcsname{1900.0}
\expandafter\def\csname ppt@rev.desc.d6.uiv.p1.C.tdiag.dual-d2.N-disagree\endcsname{$+$1900.0}
\expandafter\def\csname val@rev.desc.d6.uiv.p1.C.tdiag.dual-d2.boot-resid.ci95.0\endcsname{0.000}
\expandafter\def\csname cnt@rev.desc.d6.uiv.p1.C.tdiag.dual-d2.boot-resid.ci95.0\endcsname{0}
\expandafter\def\csname pct@rev.desc.d6.uiv.p1.C.tdiag.dual-d2.boot-resid.ci95.0\endcsname{0.0}
\expandafter\def\csname ppt@rev.desc.d6.uiv.p1.C.tdiag.dual-d2.boot-resid.ci95.0\endcsname{0.0}
\expandafter\def\csname val@rev.desc.d6.uiv.p1.C.tdiag.dual-d2.boot-resid.ci95.1\endcsname{0.048}
\expandafter\def\csname cnt@rev.desc.d6.uiv.p1.C.tdiag.dual-d2.boot-resid.ci95.1\endcsname{0}
\expandafter\def\csname pct@rev.desc.d6.uiv.p1.C.tdiag.dual-d2.boot-resid.ci95.1\endcsname{4.8}
\expandafter\def\csname ppt@rev.desc.d6.uiv.p1.C.tdiag.dual-d2.boot-resid.ci95.1\endcsname{$+$4.8}
\expandafter\def\csname val@rev.desc.d6.uiv.p1.C.tdiag.dual-d2.boot-resid.n-apps\endcsname{12}
\expandafter\def\csname cnt@rev.desc.d6.uiv.p1.C.tdiag.dual-d2.boot-resid.n-apps\endcsname{12}
\expandafter\def\csname pct@rev.desc.d6.uiv.p1.C.tdiag.dual-d2.boot-resid.n-apps\endcsname{1200.0}
\expandafter\def\csname ppt@rev.desc.d6.uiv.p1.C.tdiag.dual-d2.boot-resid.n-apps\endcsname{$+$1200.0}
\expandafter\def\csname val@rev.desc.d6.uiv.p1.C.tdiag.dual-d2.boot-resid.n-boot-valid\endcsname{4000}
\expandafter\def\csname cnt@rev.desc.d6.uiv.p1.C.tdiag.dual-d2.boot-resid.n-boot-valid\endcsname{4{,}000}
\expandafter\def\csname pct@rev.desc.d6.uiv.p1.C.tdiag.dual-d2.boot-resid.n-boot-valid\endcsname{400000.0}
\expandafter\def\csname ppt@rev.desc.d6.uiv.p1.C.tdiag.dual-d2.boot-resid.n-boot-valid\endcsname{$+$400000.0}
\expandafter\def\csname val@rev.desc.d6.uiv.p1.C.tdiag.dual-d2.boot-resid.value\endcsname{0.015}
\expandafter\def\csname cnt@rev.desc.d6.uiv.p1.C.tdiag.dual-d2.boot-resid.value\endcsname{0}
\expandafter\def\csname pct@rev.desc.d6.uiv.p1.C.tdiag.dual-d2.boot-resid.value\endcsname{1.5}
\expandafter\def\csname ppt@rev.desc.d6.uiv.p1.C.tdiag.dual-d2.boot-resid.value\endcsname{$+$1.5}
\expandafter\def\csname val@rev.desc.d6.uiv.p1.C.tdiag.dual-d2.boot-share.ci95.0\endcsname{0.635}
\expandafter\def\csname cnt@rev.desc.d6.uiv.p1.C.tdiag.dual-d2.boot-share.ci95.0\endcsname{1}
\expandafter\def\csname pct@rev.desc.d6.uiv.p1.C.tdiag.dual-d2.boot-share.ci95.0\endcsname{63.5}
\expandafter\def\csname ppt@rev.desc.d6.uiv.p1.C.tdiag.dual-d2.boot-share.ci95.0\endcsname{$+$63.5}
\expandafter\def\csname val@rev.desc.d6.uiv.p1.C.tdiag.dual-d2.boot-share.ci95.1\endcsname{0.760}
\expandafter\def\csname cnt@rev.desc.d6.uiv.p1.C.tdiag.dual-d2.boot-share.ci95.1\endcsname{1}
\expandafter\def\csname pct@rev.desc.d6.uiv.p1.C.tdiag.dual-d2.boot-share.ci95.1\endcsname{76.0}
\expandafter\def\csname ppt@rev.desc.d6.uiv.p1.C.tdiag.dual-d2.boot-share.ci95.1\endcsname{$+$76.0}
\expandafter\def\csname val@rev.desc.d6.uiv.p1.C.tdiag.dual-d2.boot-share.n-apps\endcsname{12}
\expandafter\def\csname cnt@rev.desc.d6.uiv.p1.C.tdiag.dual-d2.boot-share.n-apps\endcsname{12}
\expandafter\def\csname pct@rev.desc.d6.uiv.p1.C.tdiag.dual-d2.boot-share.n-apps\endcsname{1200.0}
\expandafter\def\csname ppt@rev.desc.d6.uiv.p1.C.tdiag.dual-d2.boot-share.n-apps\endcsname{$+$1200.0}
\expandafter\def\csname val@rev.desc.d6.uiv.p1.C.tdiag.dual-d2.boot-share.n-boot-valid\endcsname{4000}
\expandafter\def\csname cnt@rev.desc.d6.uiv.p1.C.tdiag.dual-d2.boot-share.n-boot-valid\endcsname{4{,}000}
\expandafter\def\csname pct@rev.desc.d6.uiv.p1.C.tdiag.dual-d2.boot-share.n-boot-valid\endcsname{400000.0}
\expandafter\def\csname ppt@rev.desc.d6.uiv.p1.C.tdiag.dual-d2.boot-share.n-boot-valid\endcsname{$+$400000.0}
\expandafter\def\csname val@rev.desc.d6.uiv.p1.C.tdiag.dual-d2.boot-share.value\endcsname{0.698}
\expandafter\def\csname cnt@rev.desc.d6.uiv.p1.C.tdiag.dual-d2.boot-share.value\endcsname{1}
\expandafter\def\csname pct@rev.desc.d6.uiv.p1.C.tdiag.dual-d2.boot-share.value\endcsname{69.8}
\expandafter\def\csname ppt@rev.desc.d6.uiv.p1.C.tdiag.dual-d2.boot-share.value\endcsname{$+$69.8}
\expandafter\def\csname val@rev.desc.d6.uiv.p1.C.tdiag.dual-d2.n\endcsname{96}
\expandafter\def\csname cnt@rev.desc.d6.uiv.p1.C.tdiag.dual-d2.n\endcsname{96}
\expandafter\def\csname pct@rev.desc.d6.uiv.p1.C.tdiag.dual-d2.n\endcsname{9600.0}
\expandafter\def\csname ppt@rev.desc.d6.uiv.p1.C.tdiag.dual-d2.n\endcsname{$+$9600.0}
\expandafter\def\csname val@rev.desc.d6.uiv.p1.C.tdiag.dual-d2.resid-agree-stable\endcsname{0.015}
\expandafter\def\csname cnt@rev.desc.d6.uiv.p1.C.tdiag.dual-d2.resid-agree-stable\endcsname{0}
\expandafter\def\csname pct@rev.desc.d6.uiv.p1.C.tdiag.dual-d2.resid-agree-stable\endcsname{1.5}
\expandafter\def\csname ppt@rev.desc.d6.uiv.p1.C.tdiag.dual-d2.resid-agree-stable\endcsname{$+$1.5}
\expandafter\def\csname val@rev.desc.d6.uiv.p1.C.tdiag.dual-d2.resid-disagree\endcsname{0.316}
\expandafter\def\csname cnt@rev.desc.d6.uiv.p1.C.tdiag.dual-d2.resid-disagree\endcsname{0}
\expandafter\def\csname pct@rev.desc.d6.uiv.p1.C.tdiag.dual-d2.resid-disagree\endcsname{31.6}
\expandafter\def\csname ppt@rev.desc.d6.uiv.p1.C.tdiag.dual-d2.resid-disagree\endcsname{$+$31.6}
\expandafter\def\csname val@rev.desc.d6.uiv.p1.C.tdiag.dual-d2.share-agree-stable\endcsname{0.698}
\expandafter\def\csname cnt@rev.desc.d6.uiv.p1.C.tdiag.dual-d2.share-agree-stable\endcsname{1}
\expandafter\def\csname pct@rev.desc.d6.uiv.p1.C.tdiag.dual-d2.share-agree-stable\endcsname{69.8}
\expandafter\def\csname ppt@rev.desc.d6.uiv.p1.C.tdiag.dual-d2.share-agree-stable\endcsname{$+$69.8}
\expandafter\def\csname val@rev.desc.d6.uiv.p1.C.tdiag.grid2-d3.N-agree-stable\endcsname{65}
\expandafter\def\csname cnt@rev.desc.d6.uiv.p1.C.tdiag.grid2-d3.N-agree-stable\endcsname{65}
\expandafter\def\csname pct@rev.desc.d6.uiv.p1.C.tdiag.grid2-d3.N-agree-stable\endcsname{6500.0}
\expandafter\def\csname ppt@rev.desc.d6.uiv.p1.C.tdiag.grid2-d3.N-agree-stable\endcsname{$+$6500.0}
\expandafter\def\csname val@rev.desc.d6.uiv.p1.C.tdiag.grid2-d3.N-disagree\endcsname{19}
\expandafter\def\csname cnt@rev.desc.d6.uiv.p1.C.tdiag.grid2-d3.N-disagree\endcsname{19}
\expandafter\def\csname pct@rev.desc.d6.uiv.p1.C.tdiag.grid2-d3.N-disagree\endcsname{1900.0}
\expandafter\def\csname ppt@rev.desc.d6.uiv.p1.C.tdiag.grid2-d3.N-disagree\endcsname{$+$1900.0}
\expandafter\def\csname val@rev.desc.d6.uiv.p1.C.tdiag.grid2-d3.boot-resid.ci95.0\endcsname{-0.016}
\expandafter\def\csname cnt@rev.desc.d6.uiv.p1.C.tdiag.grid2-d3.boot-resid.ci95.0\endcsname{0}
\expandafter\def\csname pct@rev.desc.d6.uiv.p1.C.tdiag.grid2-d3.boot-resid.ci95.0\endcsname{$-$1.6}
\expandafter\def\csname ppt@rev.desc.d6.uiv.p1.C.tdiag.grid2-d3.boot-resid.ci95.0\endcsname{$-$1.6}
\expandafter\def\csname val@rev.desc.d6.uiv.p1.C.tdiag.grid2-d3.boot-resid.ci95.1\endcsname{0.106}
\expandafter\def\csname cnt@rev.desc.d6.uiv.p1.C.tdiag.grid2-d3.boot-resid.ci95.1\endcsname{0}
\expandafter\def\csname pct@rev.desc.d6.uiv.p1.C.tdiag.grid2-d3.boot-resid.ci95.1\endcsname{10.6}
\expandafter\def\csname ppt@rev.desc.d6.uiv.p1.C.tdiag.grid2-d3.boot-resid.ci95.1\endcsname{$+$10.6}
\expandafter\def\csname val@rev.desc.d6.uiv.p1.C.tdiag.grid2-d3.boot-resid.n-apps\endcsname{12}
\expandafter\def\csname cnt@rev.desc.d6.uiv.p1.C.tdiag.grid2-d3.boot-resid.n-apps\endcsname{12}
\expandafter\def\csname pct@rev.desc.d6.uiv.p1.C.tdiag.grid2-d3.boot-resid.n-apps\endcsname{1200.0}
\expandafter\def\csname ppt@rev.desc.d6.uiv.p1.C.tdiag.grid2-d3.boot-resid.n-apps\endcsname{$+$1200.0}
\expandafter\def\csname val@rev.desc.d6.uiv.p1.C.tdiag.grid2-d3.boot-resid.n-boot-valid\endcsname{4000}
\expandafter\def\csname cnt@rev.desc.d6.uiv.p1.C.tdiag.grid2-d3.boot-resid.n-boot-valid\endcsname{4{,}000}
\expandafter\def\csname pct@rev.desc.d6.uiv.p1.C.tdiag.grid2-d3.boot-resid.n-boot-valid\endcsname{400000.0}
\expandafter\def\csname ppt@rev.desc.d6.uiv.p1.C.tdiag.grid2-d3.boot-resid.n-boot-valid\endcsname{$+$400000.0}
\expandafter\def\csname val@rev.desc.d6.uiv.p1.C.tdiag.grid2-d3.boot-resid.value\endcsname{0.046}
\expandafter\def\csname cnt@rev.desc.d6.uiv.p1.C.tdiag.grid2-d3.boot-resid.value\endcsname{0}
\expandafter\def\csname pct@rev.desc.d6.uiv.p1.C.tdiag.grid2-d3.boot-resid.value\endcsname{4.6}
\expandafter\def\csname ppt@rev.desc.d6.uiv.p1.C.tdiag.grid2-d3.boot-resid.value\endcsname{$+$4.6}
\expandafter\def\csname val@rev.desc.d6.uiv.p1.C.tdiag.grid2-d3.boot-share.ci95.0\endcsname{0.583}
\expandafter\def\csname cnt@rev.desc.d6.uiv.p1.C.tdiag.grid2-d3.boot-share.ci95.0\endcsname{1}
\expandafter\def\csname pct@rev.desc.d6.uiv.p1.C.tdiag.grid2-d3.boot-share.ci95.0\endcsname{58.3}
\expandafter\def\csname ppt@rev.desc.d6.uiv.p1.C.tdiag.grid2-d3.boot-share.ci95.0\endcsname{$+$58.3}
\expandafter\def\csname val@rev.desc.d6.uiv.p1.C.tdiag.grid2-d3.boot-share.ci95.1\endcsname{0.781}
\expandafter\def\csname cnt@rev.desc.d6.uiv.p1.C.tdiag.grid2-d3.boot-share.ci95.1\endcsname{1}
\expandafter\def\csname pct@rev.desc.d6.uiv.p1.C.tdiag.grid2-d3.boot-share.ci95.1\endcsname{78.1}
\expandafter\def\csname ppt@rev.desc.d6.uiv.p1.C.tdiag.grid2-d3.boot-share.ci95.1\endcsname{$+$78.1}
\expandafter\def\csname val@rev.desc.d6.uiv.p1.C.tdiag.grid2-d3.boot-share.n-apps\endcsname{12}
\expandafter\def\csname cnt@rev.desc.d6.uiv.p1.C.tdiag.grid2-d3.boot-share.n-apps\endcsname{12}
\expandafter\def\csname pct@rev.desc.d6.uiv.p1.C.tdiag.grid2-d3.boot-share.n-apps\endcsname{1200.0}
\expandafter\def\csname ppt@rev.desc.d6.uiv.p1.C.tdiag.grid2-d3.boot-share.n-apps\endcsname{$+$1200.0}
\expandafter\def\csname val@rev.desc.d6.uiv.p1.C.tdiag.grid2-d3.boot-share.n-boot-valid\endcsname{4000}
\expandafter\def\csname cnt@rev.desc.d6.uiv.p1.C.tdiag.grid2-d3.boot-share.n-boot-valid\endcsname{4{,}000}
\expandafter\def\csname pct@rev.desc.d6.uiv.p1.C.tdiag.grid2-d3.boot-share.n-boot-valid\endcsname{400000.0}
\expandafter\def\csname ppt@rev.desc.d6.uiv.p1.C.tdiag.grid2-d3.boot-share.n-boot-valid\endcsname{$+$400000.0}
\expandafter\def\csname val@rev.desc.d6.uiv.p1.C.tdiag.grid2-d3.boot-share.value\endcsname{0.677}
\expandafter\def\csname cnt@rev.desc.d6.uiv.p1.C.tdiag.grid2-d3.boot-share.value\endcsname{1}
\expandafter\def\csname pct@rev.desc.d6.uiv.p1.C.tdiag.grid2-d3.boot-share.value\endcsname{67.7}
\expandafter\def\csname ppt@rev.desc.d6.uiv.p1.C.tdiag.grid2-d3.boot-share.value\endcsname{$+$67.7}
\expandafter\def\csname val@rev.desc.d6.uiv.p1.C.tdiag.grid2-d3.n\endcsname{96}
\expandafter\def\csname cnt@rev.desc.d6.uiv.p1.C.tdiag.grid2-d3.n\endcsname{96}
\expandafter\def\csname pct@rev.desc.d6.uiv.p1.C.tdiag.grid2-d3.n\endcsname{9600.0}
\expandafter\def\csname ppt@rev.desc.d6.uiv.p1.C.tdiag.grid2-d3.n\endcsname{$+$9600.0}
\expandafter\def\csname val@rev.desc.d6.uiv.p1.C.tdiag.grid2-d3.resid-agree-stable\endcsname{0.046}
\expandafter\def\csname cnt@rev.desc.d6.uiv.p1.C.tdiag.grid2-d3.resid-agree-stable\endcsname{0}
\expandafter\def\csname pct@rev.desc.d6.uiv.p1.C.tdiag.grid2-d3.resid-agree-stable\endcsname{4.6}
\expandafter\def\csname ppt@rev.desc.d6.uiv.p1.C.tdiag.grid2-d3.resid-agree-stable\endcsname{$+$4.6}
\expandafter\def\csname val@rev.desc.d6.uiv.p1.C.tdiag.grid2-d3.resid-disagree\endcsname{0.368}
\expandafter\def\csname cnt@rev.desc.d6.uiv.p1.C.tdiag.grid2-d3.resid-disagree\endcsname{0}
\expandafter\def\csname pct@rev.desc.d6.uiv.p1.C.tdiag.grid2-d3.resid-disagree\endcsname{36.8}
\expandafter\def\csname ppt@rev.desc.d6.uiv.p1.C.tdiag.grid2-d3.resid-disagree\endcsname{$+$36.8}
\expandafter\def\csname val@rev.desc.d6.uiv.p1.C.tdiag.grid2-d3.share-agree-stable\endcsname{0.677}
\expandafter\def\csname cnt@rev.desc.d6.uiv.p1.C.tdiag.grid2-d3.share-agree-stable\endcsname{1}
\expandafter\def\csname pct@rev.desc.d6.uiv.p1.C.tdiag.grid2-d3.share-agree-stable\endcsname{67.7}
\expandafter\def\csname ppt@rev.desc.d6.uiv.p1.C.tdiag.grid2-d3.share-agree-stable\endcsname{$+$67.7}
\expandafter\def\csname val@rev.desc.d6.uiv.p1.C.tdiag.zoomclick-d3.N-agree-stable\endcsname{77}
\expandafter\def\csname cnt@rev.desc.d6.uiv.p1.C.tdiag.zoomclick-d3.N-agree-stable\endcsname{77}
\expandafter\def\csname pct@rev.desc.d6.uiv.p1.C.tdiag.zoomclick-d3.N-agree-stable\endcsname{7700.0}
\expandafter\def\csname ppt@rev.desc.d6.uiv.p1.C.tdiag.zoomclick-d3.N-agree-stable\endcsname{$+$7700.0}
\expandafter\def\csname val@rev.desc.d6.uiv.p1.C.tdiag.zoomclick-d3.N-disagree\endcsname{12}
\expandafter\def\csname cnt@rev.desc.d6.uiv.p1.C.tdiag.zoomclick-d3.N-disagree\endcsname{12}
\expandafter\def\csname pct@rev.desc.d6.uiv.p1.C.tdiag.zoomclick-d3.N-disagree\endcsname{1200.0}
\expandafter\def\csname ppt@rev.desc.d6.uiv.p1.C.tdiag.zoomclick-d3.N-disagree\endcsname{$+$1200.0}
\expandafter\def\csname val@rev.desc.d6.uiv.p1.C.tdiag.zoomclick-d3.boot-resid.ci95.0\endcsname{-0.053}
\expandafter\def\csname cnt@rev.desc.d6.uiv.p1.C.tdiag.zoomclick-d3.boot-resid.ci95.0\endcsname{0}
\expandafter\def\csname pct@rev.desc.d6.uiv.p1.C.tdiag.zoomclick-d3.boot-resid.ci95.0\endcsname{$-$5.3}
\expandafter\def\csname ppt@rev.desc.d6.uiv.p1.C.tdiag.zoomclick-d3.boot-resid.ci95.0\endcsname{$-$5.3}
\expandafter\def\csname val@rev.desc.d6.uiv.p1.C.tdiag.zoomclick-d3.boot-resid.ci95.1\endcsname{0.048}
\expandafter\def\csname cnt@rev.desc.d6.uiv.p1.C.tdiag.zoomclick-d3.boot-resid.ci95.1\endcsname{0}
\expandafter\def\csname pct@rev.desc.d6.uiv.p1.C.tdiag.zoomclick-d3.boot-resid.ci95.1\endcsname{4.8}
\expandafter\def\csname ppt@rev.desc.d6.uiv.p1.C.tdiag.zoomclick-d3.boot-resid.ci95.1\endcsname{$+$4.8}
\expandafter\def\csname val@rev.desc.d6.uiv.p1.C.tdiag.zoomclick-d3.boot-resid.n-apps\endcsname{12}
\expandafter\def\csname cnt@rev.desc.d6.uiv.p1.C.tdiag.zoomclick-d3.boot-resid.n-apps\endcsname{12}
\expandafter\def\csname pct@rev.desc.d6.uiv.p1.C.tdiag.zoomclick-d3.boot-resid.n-apps\endcsname{1200.0}
\expandafter\def\csname ppt@rev.desc.d6.uiv.p1.C.tdiag.zoomclick-d3.boot-resid.n-apps\endcsname{$+$1200.0}
\expandafter\def\csname val@rev.desc.d6.uiv.p1.C.tdiag.zoomclick-d3.boot-resid.n-boot-valid\endcsname{4000}
\expandafter\def\csname cnt@rev.desc.d6.uiv.p1.C.tdiag.zoomclick-d3.boot-resid.n-boot-valid\endcsname{4{,}000}
\expandafter\def\csname pct@rev.desc.d6.uiv.p1.C.tdiag.zoomclick-d3.boot-resid.n-boot-valid\endcsname{400000.0}
\expandafter\def\csname ppt@rev.desc.d6.uiv.p1.C.tdiag.zoomclick-d3.boot-resid.n-boot-valid\endcsname{$+$400000.0}
\expandafter\def\csname val@rev.desc.d6.uiv.p1.C.tdiag.zoomclick-d3.boot-resid.value\endcsname{0.000}
\expandafter\def\csname cnt@rev.desc.d6.uiv.p1.C.tdiag.zoomclick-d3.boot-resid.value\endcsname{0}
\expandafter\def\csname pct@rev.desc.d6.uiv.p1.C.tdiag.zoomclick-d3.boot-resid.value\endcsname{0.0}
\expandafter\def\csname ppt@rev.desc.d6.uiv.p1.C.tdiag.zoomclick-d3.boot-resid.value\endcsname{0.0}
\expandafter\def\csname val@rev.desc.d6.uiv.p1.C.tdiag.zoomclick-d3.boot-share.ci95.0\endcsname{0.719}
\expandafter\def\csname cnt@rev.desc.d6.uiv.p1.C.tdiag.zoomclick-d3.boot-share.ci95.0\endcsname{1}
\expandafter\def\csname pct@rev.desc.d6.uiv.p1.C.tdiag.zoomclick-d3.boot-share.ci95.0\endcsname{71.9}
\expandafter\def\csname ppt@rev.desc.d6.uiv.p1.C.tdiag.zoomclick-d3.boot-share.ci95.0\endcsname{$+$71.9}
\expandafter\def\csname val@rev.desc.d6.uiv.p1.C.tdiag.zoomclick-d3.boot-share.ci95.1\endcsname{0.885}
\expandafter\def\csname cnt@rev.desc.d6.uiv.p1.C.tdiag.zoomclick-d3.boot-share.ci95.1\endcsname{1}
\expandafter\def\csname pct@rev.desc.d6.uiv.p1.C.tdiag.zoomclick-d3.boot-share.ci95.1\endcsname{88.5}
\expandafter\def\csname ppt@rev.desc.d6.uiv.p1.C.tdiag.zoomclick-d3.boot-share.ci95.1\endcsname{$+$88.5}
\expandafter\def\csname val@rev.desc.d6.uiv.p1.C.tdiag.zoomclick-d3.boot-share.n-apps\endcsname{12}
\expandafter\def\csname cnt@rev.desc.d6.uiv.p1.C.tdiag.zoomclick-d3.boot-share.n-apps\endcsname{12}
\expandafter\def\csname pct@rev.desc.d6.uiv.p1.C.tdiag.zoomclick-d3.boot-share.n-apps\endcsname{1200.0}
\expandafter\def\csname ppt@rev.desc.d6.uiv.p1.C.tdiag.zoomclick-d3.boot-share.n-apps\endcsname{$+$1200.0}
\expandafter\def\csname val@rev.desc.d6.uiv.p1.C.tdiag.zoomclick-d3.boot-share.n-boot-valid\endcsname{4000}
\expandafter\def\csname cnt@rev.desc.d6.uiv.p1.C.tdiag.zoomclick-d3.boot-share.n-boot-valid\endcsname{4{,}000}
\expandafter\def\csname pct@rev.desc.d6.uiv.p1.C.tdiag.zoomclick-d3.boot-share.n-boot-valid\endcsname{400000.0}
\expandafter\def\csname ppt@rev.desc.d6.uiv.p1.C.tdiag.zoomclick-d3.boot-share.n-boot-valid\endcsname{$+$400000.0}
\expandafter\def\csname val@rev.desc.d6.uiv.p1.C.tdiag.zoomclick-d3.boot-share.value\endcsname{0.802}
\expandafter\def\csname cnt@rev.desc.d6.uiv.p1.C.tdiag.zoomclick-d3.boot-share.value\endcsname{1}
\expandafter\def\csname pct@rev.desc.d6.uiv.p1.C.tdiag.zoomclick-d3.boot-share.value\endcsname{80.2}
\expandafter\def\csname ppt@rev.desc.d6.uiv.p1.C.tdiag.zoomclick-d3.boot-share.value\endcsname{$+$80.2}
\expandafter\def\csname val@rev.desc.d6.uiv.p1.C.tdiag.zoomclick-d3.n\endcsname{96}
\expandafter\def\csname cnt@rev.desc.d6.uiv.p1.C.tdiag.zoomclick-d3.n\endcsname{96}
\expandafter\def\csname pct@rev.desc.d6.uiv.p1.C.tdiag.zoomclick-d3.n\endcsname{9600.0}
\expandafter\def\csname ppt@rev.desc.d6.uiv.p1.C.tdiag.zoomclick-d3.n\endcsname{$+$9600.0}
\expandafter\def\csname val@rev.desc.d6.uiv.p1.C.tdiag.zoomclick-d3.resid-agree-stable\endcsname{0.000}
\expandafter\def\csname cnt@rev.desc.d6.uiv.p1.C.tdiag.zoomclick-d3.resid-agree-stable\endcsname{0}
\expandafter\def\csname pct@rev.desc.d6.uiv.p1.C.tdiag.zoomclick-d3.resid-agree-stable\endcsname{0.0}
\expandafter\def\csname ppt@rev.desc.d6.uiv.p1.C.tdiag.zoomclick-d3.resid-agree-stable\endcsname{0.0}
\expandafter\def\csname val@rev.desc.d6.uiv.p1.C.tdiag.zoomclick-d3.resid-disagree\endcsname{0.083}
\expandafter\def\csname cnt@rev.desc.d6.uiv.p1.C.tdiag.zoomclick-d3.resid-disagree\endcsname{0}
\expandafter\def\csname pct@rev.desc.d6.uiv.p1.C.tdiag.zoomclick-d3.resid-disagree\endcsname{8.3}
\expandafter\def\csname ppt@rev.desc.d6.uiv.p1.C.tdiag.zoomclick-d3.resid-disagree\endcsname{$+$8.3}
\expandafter\def\csname val@rev.desc.d6.uiv.p1.C.tdiag.zoomclick-d3.share-agree-stable\endcsname{0.802}
\expandafter\def\csname cnt@rev.desc.d6.uiv.p1.C.tdiag.zoomclick-d3.share-agree-stable\endcsname{1}
\expandafter\def\csname pct@rev.desc.d6.uiv.p1.C.tdiag.zoomclick-d3.share-agree-stable\endcsname{80.2}
\expandafter\def\csname ppt@rev.desc.d6.uiv.p1.C.tdiag.zoomclick-d3.share-agree-stable\endcsname{$+$80.2}
\expandafter\def\csname val@rev.desc.d6.uiv.p2.C.box.catalog-pooled.N-agree-stable\endcsname{275}
\expandafter\def\csname cnt@rev.desc.d6.uiv.p2.C.box.catalog-pooled.N-agree-stable\endcsname{275}
\expandafter\def\csname pct@rev.desc.d6.uiv.p2.C.box.catalog-pooled.N-agree-stable\endcsname{27500.0}
\expandafter\def\csname ppt@rev.desc.d6.uiv.p2.C.box.catalog-pooled.N-agree-stable\endcsname{$+$27500.0}
\expandafter\def\csname val@rev.desc.d6.uiv.p2.C.box.catalog-pooled.N-disagree\endcsname{103}
\expandafter\def\csname cnt@rev.desc.d6.uiv.p2.C.box.catalog-pooled.N-disagree\endcsname{103}
\expandafter\def\csname pct@rev.desc.d6.uiv.p2.C.box.catalog-pooled.N-disagree\endcsname{10300.0}
\expandafter\def\csname ppt@rev.desc.d6.uiv.p2.C.box.catalog-pooled.N-disagree\endcsname{$+$10300.0}
\expandafter\def\csname val@rev.desc.d6.uiv.p2.C.box.catalog-pooled.boot-resid.ci95.0\endcsname{-0.058}
\expandafter\def\csname cnt@rev.desc.d6.uiv.p2.C.box.catalog-pooled.boot-resid.ci95.0\endcsname{0}
\expandafter\def\csname pct@rev.desc.d6.uiv.p2.C.box.catalog-pooled.boot-resid.ci95.0\endcsname{$-$5.8}
\expandafter\def\csname ppt@rev.desc.d6.uiv.p2.C.box.catalog-pooled.boot-resid.ci95.0\endcsname{$-$5.8}
\expandafter\def\csname val@rev.desc.d6.uiv.p2.C.box.catalog-pooled.boot-resid.ci95.1\endcsname{0.000}
\expandafter\def\csname cnt@rev.desc.d6.uiv.p2.C.box.catalog-pooled.boot-resid.ci95.1\endcsname{0}
\expandafter\def\csname pct@rev.desc.d6.uiv.p2.C.box.catalog-pooled.boot-resid.ci95.1\endcsname{0.0}
\expandafter\def\csname ppt@rev.desc.d6.uiv.p2.C.box.catalog-pooled.boot-resid.ci95.1\endcsname{0.0}
\expandafter\def\csname val@rev.desc.d6.uiv.p2.C.box.catalog-pooled.boot-resid.n-apps\endcsname{12}
\expandafter\def\csname cnt@rev.desc.d6.uiv.p2.C.box.catalog-pooled.boot-resid.n-apps\endcsname{12}
\expandafter\def\csname pct@rev.desc.d6.uiv.p2.C.box.catalog-pooled.boot-resid.n-apps\endcsname{1200.0}
\expandafter\def\csname ppt@rev.desc.d6.uiv.p2.C.box.catalog-pooled.boot-resid.n-apps\endcsname{$+$1200.0}
\expandafter\def\csname val@rev.desc.d6.uiv.p2.C.box.catalog-pooled.boot-resid.n-boot-valid\endcsname{4000}
\expandafter\def\csname cnt@rev.desc.d6.uiv.p2.C.box.catalog-pooled.boot-resid.n-boot-valid\endcsname{4{,}000}
\expandafter\def\csname pct@rev.desc.d6.uiv.p2.C.box.catalog-pooled.boot-resid.n-boot-valid\endcsname{400000.0}
\expandafter\def\csname ppt@rev.desc.d6.uiv.p2.C.box.catalog-pooled.boot-resid.n-boot-valid\endcsname{$+$400000.0}
\expandafter\def\csname val@rev.desc.d6.uiv.p2.C.box.catalog-pooled.boot-resid.value\endcsname{-0.025}
\expandafter\def\csname cnt@rev.desc.d6.uiv.p2.C.box.catalog-pooled.boot-resid.value\endcsname{0}
\expandafter\def\csname pct@rev.desc.d6.uiv.p2.C.box.catalog-pooled.boot-resid.value\endcsname{$-$2.5}
\expandafter\def\csname ppt@rev.desc.d6.uiv.p2.C.box.catalog-pooled.boot-resid.value\endcsname{$-$2.5}
\expandafter\def\csname val@rev.desc.d6.uiv.p2.C.box.catalog-pooled.boot-share.ci95.0\endcsname{0.646}
\expandafter\def\csname cnt@rev.desc.d6.uiv.p2.C.box.catalog-pooled.boot-share.ci95.0\endcsname{1}
\expandafter\def\csname pct@rev.desc.d6.uiv.p2.C.box.catalog-pooled.boot-share.ci95.0\endcsname{64.6}
\expandafter\def\csname ppt@rev.desc.d6.uiv.p2.C.box.catalog-pooled.boot-share.ci95.0\endcsname{$+$64.6}
\expandafter\def\csname val@rev.desc.d6.uiv.p2.C.box.catalog-pooled.boot-share.ci95.1\endcsname{0.799}
\expandafter\def\csname cnt@rev.desc.d6.uiv.p2.C.box.catalog-pooled.boot-share.ci95.1\endcsname{1}
\expandafter\def\csname pct@rev.desc.d6.uiv.p2.C.box.catalog-pooled.boot-share.ci95.1\endcsname{79.9}
\expandafter\def\csname ppt@rev.desc.d6.uiv.p2.C.box.catalog-pooled.boot-share.ci95.1\endcsname{$+$79.9}
\expandafter\def\csname val@rev.desc.d6.uiv.p2.C.box.catalog-pooled.boot-share.n-apps\endcsname{12}
\expandafter\def\csname cnt@rev.desc.d6.uiv.p2.C.box.catalog-pooled.boot-share.n-apps\endcsname{12}
\expandafter\def\csname pct@rev.desc.d6.uiv.p2.C.box.catalog-pooled.boot-share.n-apps\endcsname{1200.0}
\expandafter\def\csname ppt@rev.desc.d6.uiv.p2.C.box.catalog-pooled.boot-share.n-apps\endcsname{$+$1200.0}
\expandafter\def\csname val@rev.desc.d6.uiv.p2.C.box.catalog-pooled.boot-share.n-boot-valid\endcsname{4000}
\expandafter\def\csname cnt@rev.desc.d6.uiv.p2.C.box.catalog-pooled.boot-share.n-boot-valid\endcsname{4{,}000}
\expandafter\def\csname pct@rev.desc.d6.uiv.p2.C.box.catalog-pooled.boot-share.n-boot-valid\endcsname{400000.0}
\expandafter\def\csname ppt@rev.desc.d6.uiv.p2.C.box.catalog-pooled.boot-share.n-boot-valid\endcsname{$+$400000.0}
\expandafter\def\csname val@rev.desc.d6.uiv.p2.C.box.catalog-pooled.boot-share.value\endcsname{0.716}
\expandafter\def\csname cnt@rev.desc.d6.uiv.p2.C.box.catalog-pooled.boot-share.value\endcsname{1}
\expandafter\def\csname pct@rev.desc.d6.uiv.p2.C.box.catalog-pooled.boot-share.value\endcsname{71.6}
\expandafter\def\csname ppt@rev.desc.d6.uiv.p2.C.box.catalog-pooled.boot-share.value\endcsname{$+$71.6}
\expandafter\def\csname val@rev.desc.d6.uiv.p2.C.box.catalog-pooled.gap\endcsname{0.152}
\expandafter\def\csname cnt@rev.desc.d6.uiv.p2.C.box.catalog-pooled.gap\endcsname{0}
\expandafter\def\csname pct@rev.desc.d6.uiv.p2.C.box.catalog-pooled.gap\endcsname{15.2}
\expandafter\def\csname ppt@rev.desc.d6.uiv.p2.C.box.catalog-pooled.gap\endcsname{$+$15.2}
\expandafter\def\csname val@rev.desc.d6.uiv.p2.C.box.catalog-pooled.n\endcsname{384}
\expandafter\def\csname cnt@rev.desc.d6.uiv.p2.C.box.catalog-pooled.n\endcsname{384}
\expandafter\def\csname pct@rev.desc.d6.uiv.p2.C.box.catalog-pooled.n\endcsname{38400.0}
\expandafter\def\csname ppt@rev.desc.d6.uiv.p2.C.box.catalog-pooled.n\endcsname{$+$38400.0}
\expandafter\def\csname val@rev.desc.d6.uiv.p2.C.box.catalog-pooled.resid-agree-stable\endcsname{-0.025}
\expandafter\def\csname cnt@rev.desc.d6.uiv.p2.C.box.catalog-pooled.resid-agree-stable\endcsname{0}
\expandafter\def\csname pct@rev.desc.d6.uiv.p2.C.box.catalog-pooled.resid-agree-stable\endcsname{$-$2.5}
\expandafter\def\csname ppt@rev.desc.d6.uiv.p2.C.box.catalog-pooled.resid-agree-stable\endcsname{$-$2.5}
\expandafter\def\csname val@rev.desc.d6.uiv.p2.C.box.catalog-pooled.resid-disagree\endcsname{0.126}
\expandafter\def\csname cnt@rev.desc.d6.uiv.p2.C.box.catalog-pooled.resid-disagree\endcsname{0}
\expandafter\def\csname pct@rev.desc.d6.uiv.p2.C.box.catalog-pooled.resid-disagree\endcsname{12.6}
\expandafter\def\csname ppt@rev.desc.d6.uiv.p2.C.box.catalog-pooled.resid-disagree\endcsname{$+$12.6}
\expandafter\def\csname val@rev.desc.d6.uiv.p2.C.box.catalog-pooled.share-agree-stable\endcsname{0.716}
\expandafter\def\csname cnt@rev.desc.d6.uiv.p2.C.box.catalog-pooled.share-agree-stable\endcsname{1}
\expandafter\def\csname pct@rev.desc.d6.uiv.p2.C.box.catalog-pooled.share-agree-stable\endcsname{71.6}
\expandafter\def\csname ppt@rev.desc.d6.uiv.p2.C.box.catalog-pooled.share-agree-stable\endcsname{$+$71.6}
\expandafter\def\csname val@rev.desc.d6.uiv.p2.C.box.center-d1.N-agree-stable\endcsname{72}
\expandafter\def\csname cnt@rev.desc.d6.uiv.p2.C.box.center-d1.N-agree-stable\endcsname{72}
\expandafter\def\csname pct@rev.desc.d6.uiv.p2.C.box.center-d1.N-agree-stable\endcsname{7200.0}
\expandafter\def\csname ppt@rev.desc.d6.uiv.p2.C.box.center-d1.N-agree-stable\endcsname{$+$7200.0}
\expandafter\def\csname val@rev.desc.d6.uiv.p2.C.box.center-d1.N-disagree\endcsname{24}
\expandafter\def\csname cnt@rev.desc.d6.uiv.p2.C.box.center-d1.N-disagree\endcsname{24}
\expandafter\def\csname pct@rev.desc.d6.uiv.p2.C.box.center-d1.N-disagree\endcsname{2400.0}
\expandafter\def\csname ppt@rev.desc.d6.uiv.p2.C.box.center-d1.N-disagree\endcsname{$+$2400.0}
\expandafter\def\csname val@rev.desc.d6.uiv.p2.C.box.center-d1.boot-resid.ci95.0\endcsname{-0.070}
\expandafter\def\csname cnt@rev.desc.d6.uiv.p2.C.box.center-d1.boot-resid.ci95.0\endcsname{0}
\expandafter\def\csname pct@rev.desc.d6.uiv.p2.C.box.center-d1.boot-resid.ci95.0\endcsname{$-$7.0}
\expandafter\def\csname ppt@rev.desc.d6.uiv.p2.C.box.center-d1.boot-resid.ci95.0\endcsname{$-$7.0}
\expandafter\def\csname val@rev.desc.d6.uiv.p2.C.box.center-d1.boot-resid.ci95.1\endcsname{0.000}
\expandafter\def\csname cnt@rev.desc.d6.uiv.p2.C.box.center-d1.boot-resid.ci95.1\endcsname{0}
\expandafter\def\csname pct@rev.desc.d6.uiv.p2.C.box.center-d1.boot-resid.ci95.1\endcsname{0.0}
\expandafter\def\csname ppt@rev.desc.d6.uiv.p2.C.box.center-d1.boot-resid.ci95.1\endcsname{0.0}
\expandafter\def\csname val@rev.desc.d6.uiv.p2.C.box.center-d1.boot-resid.n-apps\endcsname{12}
\expandafter\def\csname cnt@rev.desc.d6.uiv.p2.C.box.center-d1.boot-resid.n-apps\endcsname{12}
\expandafter\def\csname pct@rev.desc.d6.uiv.p2.C.box.center-d1.boot-resid.n-apps\endcsname{1200.0}
\expandafter\def\csname ppt@rev.desc.d6.uiv.p2.C.box.center-d1.boot-resid.n-apps\endcsname{$+$1200.0}
\expandafter\def\csname val@rev.desc.d6.uiv.p2.C.box.center-d1.boot-resid.n-boot-valid\endcsname{4000}
\expandafter\def\csname cnt@rev.desc.d6.uiv.p2.C.box.center-d1.boot-resid.n-boot-valid\endcsname{4{,}000}
\expandafter\def\csname pct@rev.desc.d6.uiv.p2.C.box.center-d1.boot-resid.n-boot-valid\endcsname{400000.0}
\expandafter\def\csname ppt@rev.desc.d6.uiv.p2.C.box.center-d1.boot-resid.n-boot-valid\endcsname{$+$400000.0}
\expandafter\def\csname val@rev.desc.d6.uiv.p2.C.box.center-d1.boot-resid.value\endcsname{-0.028}
\expandafter\def\csname cnt@rev.desc.d6.uiv.p2.C.box.center-d1.boot-resid.value\endcsname{0}
\expandafter\def\csname pct@rev.desc.d6.uiv.p2.C.box.center-d1.boot-resid.value\endcsname{$-$2.8}
\expandafter\def\csname ppt@rev.desc.d6.uiv.p2.C.box.center-d1.boot-resid.value\endcsname{$-$2.8}
\expandafter\def\csname val@rev.desc.d6.uiv.p2.C.box.center-d1.boot-share.ci95.0\endcsname{0.677}
\expandafter\def\csname cnt@rev.desc.d6.uiv.p2.C.box.center-d1.boot-share.ci95.0\endcsname{1}
\expandafter\def\csname pct@rev.desc.d6.uiv.p2.C.box.center-d1.boot-share.ci95.0\endcsname{67.7}
\expandafter\def\csname ppt@rev.desc.d6.uiv.p2.C.box.center-d1.boot-share.ci95.0\endcsname{$+$67.7}
\expandafter\def\csname val@rev.desc.d6.uiv.p2.C.box.center-d1.boot-share.ci95.1\endcsname{0.833}
\expandafter\def\csname cnt@rev.desc.d6.uiv.p2.C.box.center-d1.boot-share.ci95.1\endcsname{1}
\expandafter\def\csname pct@rev.desc.d6.uiv.p2.C.box.center-d1.boot-share.ci95.1\endcsname{83.3}
\expandafter\def\csname ppt@rev.desc.d6.uiv.p2.C.box.center-d1.boot-share.ci95.1\endcsname{$+$83.3}
\expandafter\def\csname val@rev.desc.d6.uiv.p2.C.box.center-d1.boot-share.n-apps\endcsname{12}
\expandafter\def\csname cnt@rev.desc.d6.uiv.p2.C.box.center-d1.boot-share.n-apps\endcsname{12}
\expandafter\def\csname pct@rev.desc.d6.uiv.p2.C.box.center-d1.boot-share.n-apps\endcsname{1200.0}
\expandafter\def\csname ppt@rev.desc.d6.uiv.p2.C.box.center-d1.boot-share.n-apps\endcsname{$+$1200.0}
\expandafter\def\csname val@rev.desc.d6.uiv.p2.C.box.center-d1.boot-share.n-boot-valid\endcsname{4000}
\expandafter\def\csname cnt@rev.desc.d6.uiv.p2.C.box.center-d1.boot-share.n-boot-valid\endcsname{4{,}000}
\expandafter\def\csname pct@rev.desc.d6.uiv.p2.C.box.center-d1.boot-share.n-boot-valid\endcsname{400000.0}
\expandafter\def\csname ppt@rev.desc.d6.uiv.p2.C.box.center-d1.boot-share.n-boot-valid\endcsname{$+$400000.0}
\expandafter\def\csname val@rev.desc.d6.uiv.p2.C.box.center-d1.boot-share.value\endcsname{0.750}
\expandafter\def\csname cnt@rev.desc.d6.uiv.p2.C.box.center-d1.boot-share.value\endcsname{1}
\expandafter\def\csname pct@rev.desc.d6.uiv.p2.C.box.center-d1.boot-share.value\endcsname{75.0}
\expandafter\def\csname ppt@rev.desc.d6.uiv.p2.C.box.center-d1.boot-share.value\endcsname{$+$75.0}
\expandafter\def\csname val@rev.desc.d6.uiv.p2.C.box.center-d1.n\endcsname{96}
\expandafter\def\csname cnt@rev.desc.d6.uiv.p2.C.box.center-d1.n\endcsname{96}
\expandafter\def\csname pct@rev.desc.d6.uiv.p2.C.box.center-d1.n\endcsname{9600.0}
\expandafter\def\csname ppt@rev.desc.d6.uiv.p2.C.box.center-d1.n\endcsname{$+$9600.0}
\expandafter\def\csname val@rev.desc.d6.uiv.p2.C.box.center-d1.resid-agree-stable\endcsname{-0.028}
\expandafter\def\csname cnt@rev.desc.d6.uiv.p2.C.box.center-d1.resid-agree-stable\endcsname{0}
\expandafter\def\csname pct@rev.desc.d6.uiv.p2.C.box.center-d1.resid-agree-stable\endcsname{$-$2.8}
\expandafter\def\csname ppt@rev.desc.d6.uiv.p2.C.box.center-d1.resid-agree-stable\endcsname{$-$2.8}
\expandafter\def\csname val@rev.desc.d6.uiv.p2.C.box.center-d1.resid-disagree\endcsname{0.125}
\expandafter\def\csname cnt@rev.desc.d6.uiv.p2.C.box.center-d1.resid-disagree\endcsname{0}
\expandafter\def\csname pct@rev.desc.d6.uiv.p2.C.box.center-d1.resid-disagree\endcsname{12.5}
\expandafter\def\csname ppt@rev.desc.d6.uiv.p2.C.box.center-d1.resid-disagree\endcsname{$+$12.5}
\expandafter\def\csname val@rev.desc.d6.uiv.p2.C.box.center-d1.share-agree-stable\endcsname{0.750}
\expandafter\def\csname cnt@rev.desc.d6.uiv.p2.C.box.center-d1.share-agree-stable\endcsname{1}
\expandafter\def\csname pct@rev.desc.d6.uiv.p2.C.box.center-d1.share-agree-stable\endcsname{75.0}
\expandafter\def\csname ppt@rev.desc.d6.uiv.p2.C.box.center-d1.share-agree-stable\endcsname{$+$75.0}
\expandafter\def\csname val@rev.desc.d6.uiv.p2.C.box.context-d1.N-agree-stable\endcsname{69}
\expandafter\def\csname cnt@rev.desc.d6.uiv.p2.C.box.context-d1.N-agree-stable\endcsname{69}
\expandafter\def\csname pct@rev.desc.d6.uiv.p2.C.box.context-d1.N-agree-stable\endcsname{6900.0}
\expandafter\def\csname ppt@rev.desc.d6.uiv.p2.C.box.context-d1.N-agree-stable\endcsname{$+$6900.0}
\expandafter\def\csname val@rev.desc.d6.uiv.p2.C.box.context-d1.N-disagree\endcsname{25}
\expandafter\def\csname cnt@rev.desc.d6.uiv.p2.C.box.context-d1.N-disagree\endcsname{25}
\expandafter\def\csname pct@rev.desc.d6.uiv.p2.C.box.context-d1.N-disagree\endcsname{2500.0}
\expandafter\def\csname ppt@rev.desc.d6.uiv.p2.C.box.context-d1.N-disagree\endcsname{$+$2500.0}
\expandafter\def\csname val@rev.desc.d6.uiv.p2.C.box.context-d1.boot-resid.ci95.0\endcsname{-0.045}
\expandafter\def\csname cnt@rev.desc.d6.uiv.p2.C.box.context-d1.boot-resid.ci95.0\endcsname{0}
\expandafter\def\csname pct@rev.desc.d6.uiv.p2.C.box.context-d1.boot-resid.ci95.0\endcsname{$-$4.5}
\expandafter\def\csname ppt@rev.desc.d6.uiv.p2.C.box.context-d1.boot-resid.ci95.0\endcsname{$-$4.5}
\expandafter\def\csname val@rev.desc.d6.uiv.p2.C.box.context-d1.boot-resid.ci95.1\endcsname{0.000}
\expandafter\def\csname cnt@rev.desc.d6.uiv.p2.C.box.context-d1.boot-resid.ci95.1\endcsname{0}
\expandafter\def\csname pct@rev.desc.d6.uiv.p2.C.box.context-d1.boot-resid.ci95.1\endcsname{0.0}
\expandafter\def\csname ppt@rev.desc.d6.uiv.p2.C.box.context-d1.boot-resid.ci95.1\endcsname{0.0}
\expandafter\def\csname val@rev.desc.d6.uiv.p2.C.box.context-d1.boot-resid.n-apps\endcsname{12}
\expandafter\def\csname cnt@rev.desc.d6.uiv.p2.C.box.context-d1.boot-resid.n-apps\endcsname{12}
\expandafter\def\csname pct@rev.desc.d6.uiv.p2.C.box.context-d1.boot-resid.n-apps\endcsname{1200.0}
\expandafter\def\csname ppt@rev.desc.d6.uiv.p2.C.box.context-d1.boot-resid.n-apps\endcsname{$+$1200.0}
\expandafter\def\csname val@rev.desc.d6.uiv.p2.C.box.context-d1.boot-resid.n-boot-valid\endcsname{4000}
\expandafter\def\csname cnt@rev.desc.d6.uiv.p2.C.box.context-d1.boot-resid.n-boot-valid\endcsname{4{,}000}
\expandafter\def\csname pct@rev.desc.d6.uiv.p2.C.box.context-d1.boot-resid.n-boot-valid\endcsname{400000.0}
\expandafter\def\csname ppt@rev.desc.d6.uiv.p2.C.box.context-d1.boot-resid.n-boot-valid\endcsname{$+$400000.0}
\expandafter\def\csname val@rev.desc.d6.uiv.p2.C.box.context-d1.boot-resid.value\endcsname{-0.014}
\expandafter\def\csname cnt@rev.desc.d6.uiv.p2.C.box.context-d1.boot-resid.value\endcsname{0}
\expandafter\def\csname pct@rev.desc.d6.uiv.p2.C.box.context-d1.boot-resid.value\endcsname{$-$1.4}
\expandafter\def\csname ppt@rev.desc.d6.uiv.p2.C.box.context-d1.boot-resid.value\endcsname{$-$1.4}
\expandafter\def\csname val@rev.desc.d6.uiv.p2.C.box.context-d1.boot-share.ci95.0\endcsname{0.646}
\expandafter\def\csname cnt@rev.desc.d6.uiv.p2.C.box.context-d1.boot-share.ci95.0\endcsname{1}
\expandafter\def\csname pct@rev.desc.d6.uiv.p2.C.box.context-d1.boot-share.ci95.0\endcsname{64.6}
\expandafter\def\csname ppt@rev.desc.d6.uiv.p2.C.box.context-d1.boot-share.ci95.0\endcsname{$+$64.6}
\expandafter\def\csname val@rev.desc.d6.uiv.p2.C.box.context-d1.boot-share.ci95.1\endcsname{0.802}
\expandafter\def\csname cnt@rev.desc.d6.uiv.p2.C.box.context-d1.boot-share.ci95.1\endcsname{1}
\expandafter\def\csname pct@rev.desc.d6.uiv.p2.C.box.context-d1.boot-share.ci95.1\endcsname{80.2}
\expandafter\def\csname ppt@rev.desc.d6.uiv.p2.C.box.context-d1.boot-share.ci95.1\endcsname{$+$80.2}
\expandafter\def\csname val@rev.desc.d6.uiv.p2.C.box.context-d1.boot-share.n-apps\endcsname{12}
\expandafter\def\csname cnt@rev.desc.d6.uiv.p2.C.box.context-d1.boot-share.n-apps\endcsname{12}
\expandafter\def\csname pct@rev.desc.d6.uiv.p2.C.box.context-d1.boot-share.n-apps\endcsname{1200.0}
\expandafter\def\csname ppt@rev.desc.d6.uiv.p2.C.box.context-d1.boot-share.n-apps\endcsname{$+$1200.0}
\expandafter\def\csname val@rev.desc.d6.uiv.p2.C.box.context-d1.boot-share.n-boot-valid\endcsname{4000}
\expandafter\def\csname cnt@rev.desc.d6.uiv.p2.C.box.context-d1.boot-share.n-boot-valid\endcsname{4{,}000}
\expandafter\def\csname pct@rev.desc.d6.uiv.p2.C.box.context-d1.boot-share.n-boot-valid\endcsname{400000.0}
\expandafter\def\csname ppt@rev.desc.d6.uiv.p2.C.box.context-d1.boot-share.n-boot-valid\endcsname{$+$400000.0}
\expandafter\def\csname val@rev.desc.d6.uiv.p2.C.box.context-d1.boot-share.value\endcsname{0.719}
\expandafter\def\csname cnt@rev.desc.d6.uiv.p2.C.box.context-d1.boot-share.value\endcsname{1}
\expandafter\def\csname pct@rev.desc.d6.uiv.p2.C.box.context-d1.boot-share.value\endcsname{71.9}
\expandafter\def\csname ppt@rev.desc.d6.uiv.p2.C.box.context-d1.boot-share.value\endcsname{$+$71.9}
\expandafter\def\csname val@rev.desc.d6.uiv.p2.C.box.context-d1.n\endcsname{96}
\expandafter\def\csname cnt@rev.desc.d6.uiv.p2.C.box.context-d1.n\endcsname{96}
\expandafter\def\csname pct@rev.desc.d6.uiv.p2.C.box.context-d1.n\endcsname{9600.0}
\expandafter\def\csname ppt@rev.desc.d6.uiv.p2.C.box.context-d1.n\endcsname{$+$9600.0}
\expandafter\def\csname val@rev.desc.d6.uiv.p2.C.box.context-d1.resid-agree-stable\endcsname{-0.014}
\expandafter\def\csname cnt@rev.desc.d6.uiv.p2.C.box.context-d1.resid-agree-stable\endcsname{0}
\expandafter\def\csname pct@rev.desc.d6.uiv.p2.C.box.context-d1.resid-agree-stable\endcsname{$-$1.4}
\expandafter\def\csname ppt@rev.desc.d6.uiv.p2.C.box.context-d1.resid-agree-stable\endcsname{$-$1.4}
\expandafter\def\csname val@rev.desc.d6.uiv.p2.C.box.context-d1.resid-disagree\endcsname{0.000}
\expandafter\def\csname cnt@rev.desc.d6.uiv.p2.C.box.context-d1.resid-disagree\endcsname{0}
\expandafter\def\csname pct@rev.desc.d6.uiv.p2.C.box.context-d1.resid-disagree\endcsname{0.0}
\expandafter\def\csname ppt@rev.desc.d6.uiv.p2.C.box.context-d1.resid-disagree\endcsname{0.0}
\expandafter\def\csname val@rev.desc.d6.uiv.p2.C.box.context-d1.share-agree-stable\endcsname{0.719}
\expandafter\def\csname cnt@rev.desc.d6.uiv.p2.C.box.context-d1.share-agree-stable\endcsname{1}
\expandafter\def\csname pct@rev.desc.d6.uiv.p2.C.box.context-d1.share-agree-stable\endcsname{71.9}
\expandafter\def\csname ppt@rev.desc.d6.uiv.p2.C.box.context-d1.share-agree-stable\endcsname{$+$71.9}
\expandafter\def\csname val@rev.desc.d6.uiv.p2.C.box.dual-d2.N-agree-stable\endcsname{69}
\expandafter\def\csname cnt@rev.desc.d6.uiv.p2.C.box.dual-d2.N-agree-stable\endcsname{69}
\expandafter\def\csname pct@rev.desc.d6.uiv.p2.C.box.dual-d2.N-agree-stable\endcsname{6900.0}
\expandafter\def\csname ppt@rev.desc.d6.uiv.p2.C.box.dual-d2.N-agree-stable\endcsname{$+$6900.0}
\expandafter\def\csname val@rev.desc.d6.uiv.p2.C.box.dual-d2.N-disagree\endcsname{26}
\expandafter\def\csname cnt@rev.desc.d6.uiv.p2.C.box.dual-d2.N-disagree\endcsname{26}
\expandafter\def\csname pct@rev.desc.d6.uiv.p2.C.box.dual-d2.N-disagree\endcsname{2600.0}
\expandafter\def\csname ppt@rev.desc.d6.uiv.p2.C.box.dual-d2.N-disagree\endcsname{$+$2600.0}
\expandafter\def\csname val@rev.desc.d6.uiv.p2.C.box.dual-d2.boot-resid.ci95.0\endcsname{-0.090}
\expandafter\def\csname cnt@rev.desc.d6.uiv.p2.C.box.dual-d2.boot-resid.ci95.0\endcsname{0}
\expandafter\def\csname pct@rev.desc.d6.uiv.p2.C.box.dual-d2.boot-resid.ci95.0\endcsname{$-$9.0}
\expandafter\def\csname ppt@rev.desc.d6.uiv.p2.C.box.dual-d2.boot-resid.ci95.0\endcsname{$-$9.0}
\expandafter\def\csname val@rev.desc.d6.uiv.p2.C.box.dual-d2.boot-resid.ci95.1\endcsname{0.000}
\expandafter\def\csname cnt@rev.desc.d6.uiv.p2.C.box.dual-d2.boot-resid.ci95.1\endcsname{0}
\expandafter\def\csname pct@rev.desc.d6.uiv.p2.C.box.dual-d2.boot-resid.ci95.1\endcsname{0.0}
\expandafter\def\csname ppt@rev.desc.d6.uiv.p2.C.box.dual-d2.boot-resid.ci95.1\endcsname{0.0}
\expandafter\def\csname val@rev.desc.d6.uiv.p2.C.box.dual-d2.boot-resid.n-apps\endcsname{12}
\expandafter\def\csname cnt@rev.desc.d6.uiv.p2.C.box.dual-d2.boot-resid.n-apps\endcsname{12}
\expandafter\def\csname pct@rev.desc.d6.uiv.p2.C.box.dual-d2.boot-resid.n-apps\endcsname{1200.0}
\expandafter\def\csname ppt@rev.desc.d6.uiv.p2.C.box.dual-d2.boot-resid.n-apps\endcsname{$+$1200.0}
\expandafter\def\csname val@rev.desc.d6.uiv.p2.C.box.dual-d2.boot-resid.n-boot-valid\endcsname{4000}
\expandafter\def\csname cnt@rev.desc.d6.uiv.p2.C.box.dual-d2.boot-resid.n-boot-valid\endcsname{4{,}000}
\expandafter\def\csname pct@rev.desc.d6.uiv.p2.C.box.dual-d2.boot-resid.n-boot-valid\endcsname{400000.0}
\expandafter\def\csname ppt@rev.desc.d6.uiv.p2.C.box.dual-d2.boot-resid.n-boot-valid\endcsname{$+$400000.0}
\expandafter\def\csname val@rev.desc.d6.uiv.p2.C.box.dual-d2.boot-resid.value\endcsname{-0.043}
\expandafter\def\csname cnt@rev.desc.d6.uiv.p2.C.box.dual-d2.boot-resid.value\endcsname{0}
\expandafter\def\csname pct@rev.desc.d6.uiv.p2.C.box.dual-d2.boot-resid.value\endcsname{$-$4.3}
\expandafter\def\csname ppt@rev.desc.d6.uiv.p2.C.box.dual-d2.boot-resid.value\endcsname{$-$4.3}
\expandafter\def\csname val@rev.desc.d6.uiv.p2.C.box.dual-d2.boot-share.ci95.0\endcsname{0.635}
\expandafter\def\csname cnt@rev.desc.d6.uiv.p2.C.box.dual-d2.boot-share.ci95.0\endcsname{1}
\expandafter\def\csname pct@rev.desc.d6.uiv.p2.C.box.dual-d2.boot-share.ci95.0\endcsname{63.5}
\expandafter\def\csname ppt@rev.desc.d6.uiv.p2.C.box.dual-d2.boot-share.ci95.0\endcsname{$+$63.5}
\expandafter\def\csname val@rev.desc.d6.uiv.p2.C.box.dual-d2.boot-share.ci95.1\endcsname{0.812}
\expandafter\def\csname cnt@rev.desc.d6.uiv.p2.C.box.dual-d2.boot-share.ci95.1\endcsname{1}
\expandafter\def\csname pct@rev.desc.d6.uiv.p2.C.box.dual-d2.boot-share.ci95.1\endcsname{81.2}
\expandafter\def\csname ppt@rev.desc.d6.uiv.p2.C.box.dual-d2.boot-share.ci95.1\endcsname{$+$81.2}
\expandafter\def\csname val@rev.desc.d6.uiv.p2.C.box.dual-d2.boot-share.n-apps\endcsname{12}
\expandafter\def\csname cnt@rev.desc.d6.uiv.p2.C.box.dual-d2.boot-share.n-apps\endcsname{12}
\expandafter\def\csname pct@rev.desc.d6.uiv.p2.C.box.dual-d2.boot-share.n-apps\endcsname{1200.0}
\expandafter\def\csname ppt@rev.desc.d6.uiv.p2.C.box.dual-d2.boot-share.n-apps\endcsname{$+$1200.0}
\expandafter\def\csname val@rev.desc.d6.uiv.p2.C.box.dual-d2.boot-share.n-boot-valid\endcsname{4000}
\expandafter\def\csname cnt@rev.desc.d6.uiv.p2.C.box.dual-d2.boot-share.n-boot-valid\endcsname{4{,}000}
\expandafter\def\csname pct@rev.desc.d6.uiv.p2.C.box.dual-d2.boot-share.n-boot-valid\endcsname{400000.0}
\expandafter\def\csname ppt@rev.desc.d6.uiv.p2.C.box.dual-d2.boot-share.n-boot-valid\endcsname{$+$400000.0}
\expandafter\def\csname val@rev.desc.d6.uiv.p2.C.box.dual-d2.boot-share.value\endcsname{0.719}
\expandafter\def\csname cnt@rev.desc.d6.uiv.p2.C.box.dual-d2.boot-share.value\endcsname{1}
\expandafter\def\csname pct@rev.desc.d6.uiv.p2.C.box.dual-d2.boot-share.value\endcsname{71.9}
\expandafter\def\csname ppt@rev.desc.d6.uiv.p2.C.box.dual-d2.boot-share.value\endcsname{$+$71.9}
\expandafter\def\csname val@rev.desc.d6.uiv.p2.C.box.dual-d2.n\endcsname{96}
\expandafter\def\csname cnt@rev.desc.d6.uiv.p2.C.box.dual-d2.n\endcsname{96}
\expandafter\def\csname pct@rev.desc.d6.uiv.p2.C.box.dual-d2.n\endcsname{9600.0}
\expandafter\def\csname ppt@rev.desc.d6.uiv.p2.C.box.dual-d2.n\endcsname{$+$9600.0}
\expandafter\def\csname val@rev.desc.d6.uiv.p2.C.box.dual-d2.resid-agree-stable\endcsname{-0.043}
\expandafter\def\csname cnt@rev.desc.d6.uiv.p2.C.box.dual-d2.resid-agree-stable\endcsname{0}
\expandafter\def\csname pct@rev.desc.d6.uiv.p2.C.box.dual-d2.resid-agree-stable\endcsname{$-$4.3}
\expandafter\def\csname ppt@rev.desc.d6.uiv.p2.C.box.dual-d2.resid-agree-stable\endcsname{$-$4.3}
\expandafter\def\csname val@rev.desc.d6.uiv.p2.C.box.dual-d2.resid-disagree\endcsname{0.154}
\expandafter\def\csname cnt@rev.desc.d6.uiv.p2.C.box.dual-d2.resid-disagree\endcsname{0}
\expandafter\def\csname pct@rev.desc.d6.uiv.p2.C.box.dual-d2.resid-disagree\endcsname{15.4}
\expandafter\def\csname ppt@rev.desc.d6.uiv.p2.C.box.dual-d2.resid-disagree\endcsname{$+$15.4}
\expandafter\def\csname val@rev.desc.d6.uiv.p2.C.box.dual-d2.share-agree-stable\endcsname{0.719}
\expandafter\def\csname cnt@rev.desc.d6.uiv.p2.C.box.dual-d2.share-agree-stable\endcsname{1}
\expandafter\def\csname pct@rev.desc.d6.uiv.p2.C.box.dual-d2.share-agree-stable\endcsname{71.9}
\expandafter\def\csname ppt@rev.desc.d6.uiv.p2.C.box.dual-d2.share-agree-stable\endcsname{$+$71.9}
\expandafter\def\csname val@rev.desc.d6.uiv.p2.C.box.grid2-d3.N-agree-stable\endcsname{65}
\expandafter\def\csname cnt@rev.desc.d6.uiv.p2.C.box.grid2-d3.N-agree-stable\endcsname{65}
\expandafter\def\csname pct@rev.desc.d6.uiv.p2.C.box.grid2-d3.N-agree-stable\endcsname{6500.0}
\expandafter\def\csname ppt@rev.desc.d6.uiv.p2.C.box.grid2-d3.N-agree-stable\endcsname{$+$6500.0}
\expandafter\def\csname val@rev.desc.d6.uiv.p2.C.box.grid2-d3.N-disagree\endcsname{28}
\expandafter\def\csname cnt@rev.desc.d6.uiv.p2.C.box.grid2-d3.N-disagree\endcsname{28}
\expandafter\def\csname pct@rev.desc.d6.uiv.p2.C.box.grid2-d3.N-disagree\endcsname{2800.0}
\expandafter\def\csname ppt@rev.desc.d6.uiv.p2.C.box.grid2-d3.N-disagree\endcsname{$+$2800.0}
\expandafter\def\csname val@rev.desc.d6.uiv.p2.C.box.grid2-d3.boot-resid.ci95.0\endcsname{-0.048}
\expandafter\def\csname cnt@rev.desc.d6.uiv.p2.C.box.grid2-d3.boot-resid.ci95.0\endcsname{0}
\expandafter\def\csname pct@rev.desc.d6.uiv.p2.C.box.grid2-d3.boot-resid.ci95.0\endcsname{$-$4.8}
\expandafter\def\csname ppt@rev.desc.d6.uiv.p2.C.box.grid2-d3.boot-resid.ci95.0\endcsname{$-$4.8}
\expandafter\def\csname val@rev.desc.d6.uiv.p2.C.box.grid2-d3.boot-resid.ci95.1\endcsname{0.000}
\expandafter\def\csname cnt@rev.desc.d6.uiv.p2.C.box.grid2-d3.boot-resid.ci95.1\endcsname{0}
\expandafter\def\csname pct@rev.desc.d6.uiv.p2.C.box.grid2-d3.boot-resid.ci95.1\endcsname{0.0}
\expandafter\def\csname ppt@rev.desc.d6.uiv.p2.C.box.grid2-d3.boot-resid.ci95.1\endcsname{0.0}
\expandafter\def\csname val@rev.desc.d6.uiv.p2.C.box.grid2-d3.boot-resid.n-apps\endcsname{12}
\expandafter\def\csname cnt@rev.desc.d6.uiv.p2.C.box.grid2-d3.boot-resid.n-apps\endcsname{12}
\expandafter\def\csname pct@rev.desc.d6.uiv.p2.C.box.grid2-d3.boot-resid.n-apps\endcsname{1200.0}
\expandafter\def\csname ppt@rev.desc.d6.uiv.p2.C.box.grid2-d3.boot-resid.n-apps\endcsname{$+$1200.0}
\expandafter\def\csname val@rev.desc.d6.uiv.p2.C.box.grid2-d3.boot-resid.n-boot-valid\endcsname{4000}
\expandafter\def\csname cnt@rev.desc.d6.uiv.p2.C.box.grid2-d3.boot-resid.n-boot-valid\endcsname{4{,}000}
\expandafter\def\csname pct@rev.desc.d6.uiv.p2.C.box.grid2-d3.boot-resid.n-boot-valid\endcsname{400000.0}
\expandafter\def\csname ppt@rev.desc.d6.uiv.p2.C.box.grid2-d3.boot-resid.n-boot-valid\endcsname{$+$400000.0}
\expandafter\def\csname val@rev.desc.d6.uiv.p2.C.box.grid2-d3.boot-resid.value\endcsname{-0.015}
\expandafter\def\csname cnt@rev.desc.d6.uiv.p2.C.box.grid2-d3.boot-resid.value\endcsname{0}
\expandafter\def\csname pct@rev.desc.d6.uiv.p2.C.box.grid2-d3.boot-resid.value\endcsname{$-$1.5}
\expandafter\def\csname ppt@rev.desc.d6.uiv.p2.C.box.grid2-d3.boot-resid.value\endcsname{$-$1.5}
\expandafter\def\csname val@rev.desc.d6.uiv.p2.C.box.grid2-d3.boot-share.ci95.0\endcsname{0.594}
\expandafter\def\csname cnt@rev.desc.d6.uiv.p2.C.box.grid2-d3.boot-share.ci95.0\endcsname{1}
\expandafter\def\csname pct@rev.desc.d6.uiv.p2.C.box.grid2-d3.boot-share.ci95.0\endcsname{59.4}
\expandafter\def\csname ppt@rev.desc.d6.uiv.p2.C.box.grid2-d3.boot-share.ci95.0\endcsname{$+$59.4}
\expandafter\def\csname val@rev.desc.d6.uiv.p2.C.box.grid2-d3.boot-share.ci95.1\endcsname{0.760}
\expandafter\def\csname cnt@rev.desc.d6.uiv.p2.C.box.grid2-d3.boot-share.ci95.1\endcsname{1}
\expandafter\def\csname pct@rev.desc.d6.uiv.p2.C.box.grid2-d3.boot-share.ci95.1\endcsname{76.0}
\expandafter\def\csname ppt@rev.desc.d6.uiv.p2.C.box.grid2-d3.boot-share.ci95.1\endcsname{$+$76.0}
\expandafter\def\csname val@rev.desc.d6.uiv.p2.C.box.grid2-d3.boot-share.n-apps\endcsname{12}
\expandafter\def\csname cnt@rev.desc.d6.uiv.p2.C.box.grid2-d3.boot-share.n-apps\endcsname{12}
\expandafter\def\csname pct@rev.desc.d6.uiv.p2.C.box.grid2-d3.boot-share.n-apps\endcsname{1200.0}
\expandafter\def\csname ppt@rev.desc.d6.uiv.p2.C.box.grid2-d3.boot-share.n-apps\endcsname{$+$1200.0}
\expandafter\def\csname val@rev.desc.d6.uiv.p2.C.box.grid2-d3.boot-share.n-boot-valid\endcsname{4000}
\expandafter\def\csname cnt@rev.desc.d6.uiv.p2.C.box.grid2-d3.boot-share.n-boot-valid\endcsname{4{,}000}
\expandafter\def\csname pct@rev.desc.d6.uiv.p2.C.box.grid2-d3.boot-share.n-boot-valid\endcsname{400000.0}
\expandafter\def\csname ppt@rev.desc.d6.uiv.p2.C.box.grid2-d3.boot-share.n-boot-valid\endcsname{$+$400000.0}
\expandafter\def\csname val@rev.desc.d6.uiv.p2.C.box.grid2-d3.boot-share.value\endcsname{0.677}
\expandafter\def\csname cnt@rev.desc.d6.uiv.p2.C.box.grid2-d3.boot-share.value\endcsname{1}
\expandafter\def\csname pct@rev.desc.d6.uiv.p2.C.box.grid2-d3.boot-share.value\endcsname{67.7}
\expandafter\def\csname ppt@rev.desc.d6.uiv.p2.C.box.grid2-d3.boot-share.value\endcsname{$+$67.7}
\expandafter\def\csname val@rev.desc.d6.uiv.p2.C.box.grid2-d3.n\endcsname{96}
\expandafter\def\csname cnt@rev.desc.d6.uiv.p2.C.box.grid2-d3.n\endcsname{96}
\expandafter\def\csname pct@rev.desc.d6.uiv.p2.C.box.grid2-d3.n\endcsname{9600.0}
\expandafter\def\csname ppt@rev.desc.d6.uiv.p2.C.box.grid2-d3.n\endcsname{$+$9600.0}
\expandafter\def\csname val@rev.desc.d6.uiv.p2.C.box.grid2-d3.resid-agree-stable\endcsname{-0.015}
\expandafter\def\csname cnt@rev.desc.d6.uiv.p2.C.box.grid2-d3.resid-agree-stable\endcsname{0}
\expandafter\def\csname pct@rev.desc.d6.uiv.p2.C.box.grid2-d3.resid-agree-stable\endcsname{$-$1.5}
\expandafter\def\csname ppt@rev.desc.d6.uiv.p2.C.box.grid2-d3.resid-agree-stable\endcsname{$-$1.5}
\expandafter\def\csname val@rev.desc.d6.uiv.p2.C.box.grid2-d3.resid-disagree\endcsname{0.214}
\expandafter\def\csname cnt@rev.desc.d6.uiv.p2.C.box.grid2-d3.resid-disagree\endcsname{0}
\expandafter\def\csname pct@rev.desc.d6.uiv.p2.C.box.grid2-d3.resid-disagree\endcsname{21.4}
\expandafter\def\csname ppt@rev.desc.d6.uiv.p2.C.box.grid2-d3.resid-disagree\endcsname{$+$21.4}
\expandafter\def\csname val@rev.desc.d6.uiv.p2.C.box.grid2-d3.share-agree-stable\endcsname{0.677}
\expandafter\def\csname cnt@rev.desc.d6.uiv.p2.C.box.grid2-d3.share-agree-stable\endcsname{1}
\expandafter\def\csname pct@rev.desc.d6.uiv.p2.C.box.grid2-d3.share-agree-stable\endcsname{67.7}
\expandafter\def\csname ppt@rev.desc.d6.uiv.p2.C.box.grid2-d3.share-agree-stable\endcsname{$+$67.7}
\expandafter\def\csname val@rev.desc.d6.uiv.p2.C.box.zoomclick-d3.N-agree-stable\endcsname{70}
\expandafter\def\csname cnt@rev.desc.d6.uiv.p2.C.box.zoomclick-d3.N-agree-stable\endcsname{70}
\expandafter\def\csname pct@rev.desc.d6.uiv.p2.C.box.zoomclick-d3.N-agree-stable\endcsname{7000.0}
\expandafter\def\csname ppt@rev.desc.d6.uiv.p2.C.box.zoomclick-d3.N-agree-stable\endcsname{$+$7000.0}
\expandafter\def\csname val@rev.desc.d6.uiv.p2.C.box.zoomclick-d3.N-disagree\endcsname{26}
\expandafter\def\csname cnt@rev.desc.d6.uiv.p2.C.box.zoomclick-d3.N-disagree\endcsname{26}
\expandafter\def\csname pct@rev.desc.d6.uiv.p2.C.box.zoomclick-d3.N-disagree\endcsname{2600.0}
\expandafter\def\csname ppt@rev.desc.d6.uiv.p2.C.box.zoomclick-d3.N-disagree\endcsname{$+$2600.0}
\expandafter\def\csname val@rev.desc.d6.uiv.p2.C.box.zoomclick-d3.boot-resid.ci95.0\endcsname{-0.045}
\expandafter\def\csname cnt@rev.desc.d6.uiv.p2.C.box.zoomclick-d3.boot-resid.ci95.0\endcsname{0}
\expandafter\def\csname pct@rev.desc.d6.uiv.p2.C.box.zoomclick-d3.boot-resid.ci95.0\endcsname{$-$4.5}
\expandafter\def\csname ppt@rev.desc.d6.uiv.p2.C.box.zoomclick-d3.boot-resid.ci95.0\endcsname{$-$4.5}
\expandafter\def\csname val@rev.desc.d6.uiv.p2.C.box.zoomclick-d3.boot-resid.ci95.1\endcsname{0.000}
\expandafter\def\csname cnt@rev.desc.d6.uiv.p2.C.box.zoomclick-d3.boot-resid.ci95.1\endcsname{0}
\expandafter\def\csname pct@rev.desc.d6.uiv.p2.C.box.zoomclick-d3.boot-resid.ci95.1\endcsname{0.0}
\expandafter\def\csname ppt@rev.desc.d6.uiv.p2.C.box.zoomclick-d3.boot-resid.ci95.1\endcsname{0.0}
\expandafter\def\csname val@rev.desc.d6.uiv.p2.C.box.zoomclick-d3.boot-resid.n-apps\endcsname{12}
\expandafter\def\csname cnt@rev.desc.d6.uiv.p2.C.box.zoomclick-d3.boot-resid.n-apps\endcsname{12}
\expandafter\def\csname pct@rev.desc.d6.uiv.p2.C.box.zoomclick-d3.boot-resid.n-apps\endcsname{1200.0}
\expandafter\def\csname ppt@rev.desc.d6.uiv.p2.C.box.zoomclick-d3.boot-resid.n-apps\endcsname{$+$1200.0}
\expandafter\def\csname val@rev.desc.d6.uiv.p2.C.box.zoomclick-d3.boot-resid.n-boot-valid\endcsname{4000}
\expandafter\def\csname cnt@rev.desc.d6.uiv.p2.C.box.zoomclick-d3.boot-resid.n-boot-valid\endcsname{4{,}000}
\expandafter\def\csname pct@rev.desc.d6.uiv.p2.C.box.zoomclick-d3.boot-resid.n-boot-valid\endcsname{400000.0}
\expandafter\def\csname ppt@rev.desc.d6.uiv.p2.C.box.zoomclick-d3.boot-resid.n-boot-valid\endcsname{$+$400000.0}
\expandafter\def\csname val@rev.desc.d6.uiv.p2.C.box.zoomclick-d3.boot-resid.value\endcsname{-0.014}
\expandafter\def\csname cnt@rev.desc.d6.uiv.p2.C.box.zoomclick-d3.boot-resid.value\endcsname{0}
\expandafter\def\csname pct@rev.desc.d6.uiv.p2.C.box.zoomclick-d3.boot-resid.value\endcsname{$-$1.4}
\expandafter\def\csname ppt@rev.desc.d6.uiv.p2.C.box.zoomclick-d3.boot-resid.value\endcsname{$-$1.4}
\expandafter\def\csname val@rev.desc.d6.uiv.p2.C.box.zoomclick-d3.boot-share.ci95.0\endcsname{0.635}
\expandafter\def\csname cnt@rev.desc.d6.uiv.p2.C.box.zoomclick-d3.boot-share.ci95.0\endcsname{1}
\expandafter\def\csname pct@rev.desc.d6.uiv.p2.C.box.zoomclick-d3.boot-share.ci95.0\endcsname{63.5}
\expandafter\def\csname ppt@rev.desc.d6.uiv.p2.C.box.zoomclick-d3.boot-share.ci95.0\endcsname{$+$63.5}
\expandafter\def\csname val@rev.desc.d6.uiv.p2.C.box.zoomclick-d3.boot-share.ci95.1\endcsname{0.823}
\expandafter\def\csname cnt@rev.desc.d6.uiv.p2.C.box.zoomclick-d3.boot-share.ci95.1\endcsname{1}
\expandafter\def\csname pct@rev.desc.d6.uiv.p2.C.box.zoomclick-d3.boot-share.ci95.1\endcsname{82.3}
\expandafter\def\csname ppt@rev.desc.d6.uiv.p2.C.box.zoomclick-d3.boot-share.ci95.1\endcsname{$+$82.3}
\expandafter\def\csname val@rev.desc.d6.uiv.p2.C.box.zoomclick-d3.boot-share.n-apps\endcsname{12}
\expandafter\def\csname cnt@rev.desc.d6.uiv.p2.C.box.zoomclick-d3.boot-share.n-apps\endcsname{12}
\expandafter\def\csname pct@rev.desc.d6.uiv.p2.C.box.zoomclick-d3.boot-share.n-apps\endcsname{1200.0}
\expandafter\def\csname ppt@rev.desc.d6.uiv.p2.C.box.zoomclick-d3.boot-share.n-apps\endcsname{$+$1200.0}
\expandafter\def\csname val@rev.desc.d6.uiv.p2.C.box.zoomclick-d3.boot-share.n-boot-valid\endcsname{4000}
\expandafter\def\csname cnt@rev.desc.d6.uiv.p2.C.box.zoomclick-d3.boot-share.n-boot-valid\endcsname{4{,}000}
\expandafter\def\csname pct@rev.desc.d6.uiv.p2.C.box.zoomclick-d3.boot-share.n-boot-valid\endcsname{400000.0}
\expandafter\def\csname ppt@rev.desc.d6.uiv.p2.C.box.zoomclick-d3.boot-share.n-boot-valid\endcsname{$+$400000.0}
\expandafter\def\csname val@rev.desc.d6.uiv.p2.C.box.zoomclick-d3.boot-share.value\endcsname{0.729}
\expandafter\def\csname cnt@rev.desc.d6.uiv.p2.C.box.zoomclick-d3.boot-share.value\endcsname{1}
\expandafter\def\csname pct@rev.desc.d6.uiv.p2.C.box.zoomclick-d3.boot-share.value\endcsname{72.9}
\expandafter\def\csname ppt@rev.desc.d6.uiv.p2.C.box.zoomclick-d3.boot-share.value\endcsname{$+$72.9}
\expandafter\def\csname val@rev.desc.d6.uiv.p2.C.box.zoomclick-d3.n\endcsname{96}
\expandafter\def\csname cnt@rev.desc.d6.uiv.p2.C.box.zoomclick-d3.n\endcsname{96}
\expandafter\def\csname pct@rev.desc.d6.uiv.p2.C.box.zoomclick-d3.n\endcsname{9600.0}
\expandafter\def\csname ppt@rev.desc.d6.uiv.p2.C.box.zoomclick-d3.n\endcsname{$+$9600.0}
\expandafter\def\csname val@rev.desc.d6.uiv.p2.C.box.zoomclick-d3.resid-agree-stable\endcsname{-0.014}
\expandafter\def\csname cnt@rev.desc.d6.uiv.p2.C.box.zoomclick-d3.resid-agree-stable\endcsname{0}
\expandafter\def\csname pct@rev.desc.d6.uiv.p2.C.box.zoomclick-d3.resid-agree-stable\endcsname{$-$1.4}
\expandafter\def\csname ppt@rev.desc.d6.uiv.p2.C.box.zoomclick-d3.resid-agree-stable\endcsname{$-$1.4}
\expandafter\def\csname val@rev.desc.d6.uiv.p2.C.box.zoomclick-d3.resid-disagree\endcsname{0.192}
\expandafter\def\csname cnt@rev.desc.d6.uiv.p2.C.box.zoomclick-d3.resid-disagree\endcsname{0}
\expandafter\def\csname pct@rev.desc.d6.uiv.p2.C.box.zoomclick-d3.resid-disagree\endcsname{19.2}
\expandafter\def\csname ppt@rev.desc.d6.uiv.p2.C.box.zoomclick-d3.resid-disagree\endcsname{$+$19.2}
\expandafter\def\csname val@rev.desc.d6.uiv.p2.C.box.zoomclick-d3.share-agree-stable\endcsname{0.729}
\expandafter\def\csname cnt@rev.desc.d6.uiv.p2.C.box.zoomclick-d3.share-agree-stable\endcsname{1}
\expandafter\def\csname pct@rev.desc.d6.uiv.p2.C.box.zoomclick-d3.share-agree-stable\endcsname{72.9}
\expandafter\def\csname ppt@rev.desc.d6.uiv.p2.C.box.zoomclick-d3.share-agree-stable\endcsname{$+$72.9}
\expandafter\def\csname val@rev.desc.d6.uiv.p2.C.tdiag.catalog-pooled.N-agree-stable\endcsname{334}
\expandafter\def\csname cnt@rev.desc.d6.uiv.p2.C.tdiag.catalog-pooled.N-agree-stable\endcsname{334}
\expandafter\def\csname pct@rev.desc.d6.uiv.p2.C.tdiag.catalog-pooled.N-agree-stable\endcsname{33400.0}
\expandafter\def\csname ppt@rev.desc.d6.uiv.p2.C.tdiag.catalog-pooled.N-agree-stable\endcsname{$+$33400.0}
\expandafter\def\csname val@rev.desc.d6.uiv.p2.C.tdiag.catalog-pooled.N-disagree\endcsname{40}
\expandafter\def\csname cnt@rev.desc.d6.uiv.p2.C.tdiag.catalog-pooled.N-disagree\endcsname{40}
\expandafter\def\csname pct@rev.desc.d6.uiv.p2.C.tdiag.catalog-pooled.N-disagree\endcsname{4000.0}
\expandafter\def\csname ppt@rev.desc.d6.uiv.p2.C.tdiag.catalog-pooled.N-disagree\endcsname{$+$4000.0}
\expandafter\def\csname val@rev.desc.d6.uiv.p2.C.tdiag.catalog-pooled.boot-resid.ci95.0\endcsname{-0.028}
\expandafter\def\csname cnt@rev.desc.d6.uiv.p2.C.tdiag.catalog-pooled.boot-resid.ci95.0\endcsname{0}
\expandafter\def\csname pct@rev.desc.d6.uiv.p2.C.tdiag.catalog-pooled.boot-resid.ci95.0\endcsname{$-$2.8}
\expandafter\def\csname ppt@rev.desc.d6.uiv.p2.C.tdiag.catalog-pooled.boot-resid.ci95.0\endcsname{$-$2.8}
\expandafter\def\csname val@rev.desc.d6.uiv.p2.C.tdiag.catalog-pooled.boot-resid.ci95.1\endcsname{0.047}
\expandafter\def\csname cnt@rev.desc.d6.uiv.p2.C.tdiag.catalog-pooled.boot-resid.ci95.1\endcsname{0}
\expandafter\def\csname pct@rev.desc.d6.uiv.p2.C.tdiag.catalog-pooled.boot-resid.ci95.1\endcsname{4.7}
\expandafter\def\csname ppt@rev.desc.d6.uiv.p2.C.tdiag.catalog-pooled.boot-resid.ci95.1\endcsname{$+$4.7}
\expandafter\def\csname val@rev.desc.d6.uiv.p2.C.tdiag.catalog-pooled.boot-resid.n-apps\endcsname{12}
\expandafter\def\csname cnt@rev.desc.d6.uiv.p2.C.tdiag.catalog-pooled.boot-resid.n-apps\endcsname{12}
\expandafter\def\csname pct@rev.desc.d6.uiv.p2.C.tdiag.catalog-pooled.boot-resid.n-apps\endcsname{1200.0}
\expandafter\def\csname ppt@rev.desc.d6.uiv.p2.C.tdiag.catalog-pooled.boot-resid.n-apps\endcsname{$+$1200.0}
\expandafter\def\csname val@rev.desc.d6.uiv.p2.C.tdiag.catalog-pooled.boot-resid.n-boot-valid\endcsname{4000}
\expandafter\def\csname cnt@rev.desc.d6.uiv.p2.C.tdiag.catalog-pooled.boot-resid.n-boot-valid\endcsname{4{,}000}
\expandafter\def\csname pct@rev.desc.d6.uiv.p2.C.tdiag.catalog-pooled.boot-resid.n-boot-valid\endcsname{400000.0}
\expandafter\def\csname ppt@rev.desc.d6.uiv.p2.C.tdiag.catalog-pooled.boot-resid.n-boot-valid\endcsname{$+$400000.0}
\expandafter\def\csname val@rev.desc.d6.uiv.p2.C.tdiag.catalog-pooled.boot-resid.value\endcsname{0.009}
\expandafter\def\csname cnt@rev.desc.d6.uiv.p2.C.tdiag.catalog-pooled.boot-resid.value\endcsname{0}
\expandafter\def\csname pct@rev.desc.d6.uiv.p2.C.tdiag.catalog-pooled.boot-resid.value\endcsname{0.9}
\expandafter\def\csname ppt@rev.desc.d6.uiv.p2.C.tdiag.catalog-pooled.boot-resid.value\endcsname{$+$0.9}
\expandafter\def\csname val@rev.desc.d6.uiv.p2.C.tdiag.catalog-pooled.boot-share.ci95.0\endcsname{0.818}
\expandafter\def\csname cnt@rev.desc.d6.uiv.p2.C.tdiag.catalog-pooled.boot-share.ci95.0\endcsname{1}
\expandafter\def\csname pct@rev.desc.d6.uiv.p2.C.tdiag.catalog-pooled.boot-share.ci95.0\endcsname{81.8}
\expandafter\def\csname ppt@rev.desc.d6.uiv.p2.C.tdiag.catalog-pooled.boot-share.ci95.0\endcsname{$+$81.8}
\expandafter\def\csname val@rev.desc.d6.uiv.p2.C.tdiag.catalog-pooled.boot-share.ci95.1\endcsname{0.917}
\expandafter\def\csname cnt@rev.desc.d6.uiv.p2.C.tdiag.catalog-pooled.boot-share.ci95.1\endcsname{1}
\expandafter\def\csname pct@rev.desc.d6.uiv.p2.C.tdiag.catalog-pooled.boot-share.ci95.1\endcsname{91.7}
\expandafter\def\csname ppt@rev.desc.d6.uiv.p2.C.tdiag.catalog-pooled.boot-share.ci95.1\endcsname{$+$91.7}
\expandafter\def\csname val@rev.desc.d6.uiv.p2.C.tdiag.catalog-pooled.boot-share.n-apps\endcsname{12}
\expandafter\def\csname cnt@rev.desc.d6.uiv.p2.C.tdiag.catalog-pooled.boot-share.n-apps\endcsname{12}
\expandafter\def\csname pct@rev.desc.d6.uiv.p2.C.tdiag.catalog-pooled.boot-share.n-apps\endcsname{1200.0}
\expandafter\def\csname ppt@rev.desc.d6.uiv.p2.C.tdiag.catalog-pooled.boot-share.n-apps\endcsname{$+$1200.0}
\expandafter\def\csname val@rev.desc.d6.uiv.p2.C.tdiag.catalog-pooled.boot-share.n-boot-valid\endcsname{4000}
\expandafter\def\csname cnt@rev.desc.d6.uiv.p2.C.tdiag.catalog-pooled.boot-share.n-boot-valid\endcsname{4{,}000}
\expandafter\def\csname pct@rev.desc.d6.uiv.p2.C.tdiag.catalog-pooled.boot-share.n-boot-valid\endcsname{400000.0}
\expandafter\def\csname ppt@rev.desc.d6.uiv.p2.C.tdiag.catalog-pooled.boot-share.n-boot-valid\endcsname{$+$400000.0}
\expandafter\def\csname val@rev.desc.d6.uiv.p2.C.tdiag.catalog-pooled.boot-share.value\endcsname{0.870}
\expandafter\def\csname cnt@rev.desc.d6.uiv.p2.C.tdiag.catalog-pooled.boot-share.value\endcsname{1}
\expandafter\def\csname pct@rev.desc.d6.uiv.p2.C.tdiag.catalog-pooled.boot-share.value\endcsname{87.0}
\expandafter\def\csname ppt@rev.desc.d6.uiv.p2.C.tdiag.catalog-pooled.boot-share.value\endcsname{$+$87.0}
\expandafter\def\csname val@rev.desc.d6.uiv.p2.C.tdiag.catalog-pooled.gap\endcsname{0.016}
\expandafter\def\csname cnt@rev.desc.d6.uiv.p2.C.tdiag.catalog-pooled.gap\endcsname{0}
\expandafter\def\csname pct@rev.desc.d6.uiv.p2.C.tdiag.catalog-pooled.gap\endcsname{1.6}
\expandafter\def\csname ppt@rev.desc.d6.uiv.p2.C.tdiag.catalog-pooled.gap\endcsname{$+$1.6}
\expandafter\def\csname val@rev.desc.d6.uiv.p2.C.tdiag.catalog-pooled.n\endcsname{384}
\expandafter\def\csname cnt@rev.desc.d6.uiv.p2.C.tdiag.catalog-pooled.n\endcsname{384}
\expandafter\def\csname pct@rev.desc.d6.uiv.p2.C.tdiag.catalog-pooled.n\endcsname{38400.0}
\expandafter\def\csname ppt@rev.desc.d6.uiv.p2.C.tdiag.catalog-pooled.n\endcsname{$+$38400.0}
\expandafter\def\csname val@rev.desc.d6.uiv.p2.C.tdiag.catalog-pooled.resid-agree-stable\endcsname{0.009}
\expandafter\def\csname cnt@rev.desc.d6.uiv.p2.C.tdiag.catalog-pooled.resid-agree-stable\endcsname{0}
\expandafter\def\csname pct@rev.desc.d6.uiv.p2.C.tdiag.catalog-pooled.resid-agree-stable\endcsname{0.9}
\expandafter\def\csname ppt@rev.desc.d6.uiv.p2.C.tdiag.catalog-pooled.resid-agree-stable\endcsname{$+$0.9}
\expandafter\def\csname val@rev.desc.d6.uiv.p2.C.tdiag.catalog-pooled.resid-disagree\endcsname{0.025}
\expandafter\def\csname cnt@rev.desc.d6.uiv.p2.C.tdiag.catalog-pooled.resid-disagree\endcsname{0}
\expandafter\def\csname pct@rev.desc.d6.uiv.p2.C.tdiag.catalog-pooled.resid-disagree\endcsname{2.5}
\expandafter\def\csname ppt@rev.desc.d6.uiv.p2.C.tdiag.catalog-pooled.resid-disagree\endcsname{$+$2.5}
\expandafter\def\csname val@rev.desc.d6.uiv.p2.C.tdiag.catalog-pooled.share-agree-stable\endcsname{0.870}
\expandafter\def\csname cnt@rev.desc.d6.uiv.p2.C.tdiag.catalog-pooled.share-agree-stable\endcsname{1}
\expandafter\def\csname pct@rev.desc.d6.uiv.p2.C.tdiag.catalog-pooled.share-agree-stable\endcsname{87.0}
\expandafter\def\csname ppt@rev.desc.d6.uiv.p2.C.tdiag.catalog-pooled.share-agree-stable\endcsname{$+$87.0}
\expandafter\def\csname val@rev.desc.d6.uiv.p2.C.tdiag.center-d1.N-agree-stable\endcsname{87}
\expandafter\def\csname cnt@rev.desc.d6.uiv.p2.C.tdiag.center-d1.N-agree-stable\endcsname{87}
\expandafter\def\csname pct@rev.desc.d6.uiv.p2.C.tdiag.center-d1.N-agree-stable\endcsname{8700.0}
\expandafter\def\csname ppt@rev.desc.d6.uiv.p2.C.tdiag.center-d1.N-agree-stable\endcsname{$+$8700.0}
\expandafter\def\csname val@rev.desc.d6.uiv.p2.C.tdiag.center-d1.N-disagree\endcsname{8}
\expandafter\def\csname cnt@rev.desc.d6.uiv.p2.C.tdiag.center-d1.N-disagree\endcsname{8}
\expandafter\def\csname pct@rev.desc.d6.uiv.p2.C.tdiag.center-d1.N-disagree\endcsname{800.0}
\expandafter\def\csname ppt@rev.desc.d6.uiv.p2.C.tdiag.center-d1.N-disagree\endcsname{$+$800.0}
\expandafter\def\csname val@rev.desc.d6.uiv.p2.C.tdiag.center-d1.boot-resid.ci95.0\endcsname{-0.033}
\expandafter\def\csname cnt@rev.desc.d6.uiv.p2.C.tdiag.center-d1.boot-resid.ci95.0\endcsname{0}
\expandafter\def\csname pct@rev.desc.d6.uiv.p2.C.tdiag.center-d1.boot-resid.ci95.0\endcsname{$-$3.3}
\expandafter\def\csname ppt@rev.desc.d6.uiv.p2.C.tdiag.center-d1.boot-resid.ci95.0\endcsname{$-$3.3}
\expandafter\def\csname val@rev.desc.d6.uiv.p2.C.tdiag.center-d1.boot-resid.ci95.1\endcsname{0.035}
\expandafter\def\csname cnt@rev.desc.d6.uiv.p2.C.tdiag.center-d1.boot-resid.ci95.1\endcsname{0}
\expandafter\def\csname pct@rev.desc.d6.uiv.p2.C.tdiag.center-d1.boot-resid.ci95.1\endcsname{3.5}
\expandafter\def\csname ppt@rev.desc.d6.uiv.p2.C.tdiag.center-d1.boot-resid.ci95.1\endcsname{$+$3.5}
\expandafter\def\csname val@rev.desc.d6.uiv.p2.C.tdiag.center-d1.boot-resid.n-apps\endcsname{12}
\expandafter\def\csname cnt@rev.desc.d6.uiv.p2.C.tdiag.center-d1.boot-resid.n-apps\endcsname{12}
\expandafter\def\csname pct@rev.desc.d6.uiv.p2.C.tdiag.center-d1.boot-resid.n-apps\endcsname{1200.0}
\expandafter\def\csname ppt@rev.desc.d6.uiv.p2.C.tdiag.center-d1.boot-resid.n-apps\endcsname{$+$1200.0}
\expandafter\def\csname val@rev.desc.d6.uiv.p2.C.tdiag.center-d1.boot-resid.n-boot-valid\endcsname{4000}
\expandafter\def\csname cnt@rev.desc.d6.uiv.p2.C.tdiag.center-d1.boot-resid.n-boot-valid\endcsname{4{,}000}
\expandafter\def\csname pct@rev.desc.d6.uiv.p2.C.tdiag.center-d1.boot-resid.n-boot-valid\endcsname{400000.0}
\expandafter\def\csname ppt@rev.desc.d6.uiv.p2.C.tdiag.center-d1.boot-resid.n-boot-valid\endcsname{$+$400000.0}
\expandafter\def\csname val@rev.desc.d6.uiv.p2.C.tdiag.center-d1.boot-resid.value\endcsname{0.000}
\expandafter\def\csname cnt@rev.desc.d6.uiv.p2.C.tdiag.center-d1.boot-resid.value\endcsname{0}
\expandafter\def\csname pct@rev.desc.d6.uiv.p2.C.tdiag.center-d1.boot-resid.value\endcsname{0.0}
\expandafter\def\csname ppt@rev.desc.d6.uiv.p2.C.tdiag.center-d1.boot-resid.value\endcsname{0.0}
\expandafter\def\csname val@rev.desc.d6.uiv.p2.C.tdiag.center-d1.boot-share.ci95.0\endcsname{0.854}
\expandafter\def\csname cnt@rev.desc.d6.uiv.p2.C.tdiag.center-d1.boot-share.ci95.0\endcsname{1}
\expandafter\def\csname pct@rev.desc.d6.uiv.p2.C.tdiag.center-d1.boot-share.ci95.0\endcsname{85.4}
\expandafter\def\csname ppt@rev.desc.d6.uiv.p2.C.tdiag.center-d1.boot-share.ci95.0\endcsname{$+$85.4}
\expandafter\def\csname val@rev.desc.d6.uiv.p2.C.tdiag.center-d1.boot-share.ci95.1\endcsname{0.958}
\expandafter\def\csname cnt@rev.desc.d6.uiv.p2.C.tdiag.center-d1.boot-share.ci95.1\endcsname{1}
\expandafter\def\csname pct@rev.desc.d6.uiv.p2.C.tdiag.center-d1.boot-share.ci95.1\endcsname{95.8}
\expandafter\def\csname ppt@rev.desc.d6.uiv.p2.C.tdiag.center-d1.boot-share.ci95.1\endcsname{$+$95.8}
\expandafter\def\csname val@rev.desc.d6.uiv.p2.C.tdiag.center-d1.boot-share.n-apps\endcsname{12}
\expandafter\def\csname cnt@rev.desc.d6.uiv.p2.C.tdiag.center-d1.boot-share.n-apps\endcsname{12}
\expandafter\def\csname pct@rev.desc.d6.uiv.p2.C.tdiag.center-d1.boot-share.n-apps\endcsname{1200.0}
\expandafter\def\csname ppt@rev.desc.d6.uiv.p2.C.tdiag.center-d1.boot-share.n-apps\endcsname{$+$1200.0}
\expandafter\def\csname val@rev.desc.d6.uiv.p2.C.tdiag.center-d1.boot-share.n-boot-valid\endcsname{4000}
\expandafter\def\csname cnt@rev.desc.d6.uiv.p2.C.tdiag.center-d1.boot-share.n-boot-valid\endcsname{4{,}000}
\expandafter\def\csname pct@rev.desc.d6.uiv.p2.C.tdiag.center-d1.boot-share.n-boot-valid\endcsname{400000.0}
\expandafter\def\csname ppt@rev.desc.d6.uiv.p2.C.tdiag.center-d1.boot-share.n-boot-valid\endcsname{$+$400000.0}
\expandafter\def\csname val@rev.desc.d6.uiv.p2.C.tdiag.center-d1.boot-share.value\endcsname{0.906}
\expandafter\def\csname cnt@rev.desc.d6.uiv.p2.C.tdiag.center-d1.boot-share.value\endcsname{1}
\expandafter\def\csname pct@rev.desc.d6.uiv.p2.C.tdiag.center-d1.boot-share.value\endcsname{90.6}
\expandafter\def\csname ppt@rev.desc.d6.uiv.p2.C.tdiag.center-d1.boot-share.value\endcsname{$+$90.6}
\expandafter\def\csname val@rev.desc.d6.uiv.p2.C.tdiag.center-d1.n\endcsname{96}
\expandafter\def\csname cnt@rev.desc.d6.uiv.p2.C.tdiag.center-d1.n\endcsname{96}
\expandafter\def\csname pct@rev.desc.d6.uiv.p2.C.tdiag.center-d1.n\endcsname{9600.0}
\expandafter\def\csname ppt@rev.desc.d6.uiv.p2.C.tdiag.center-d1.n\endcsname{$+$9600.0}
\expandafter\def\csname val@rev.desc.d6.uiv.p2.C.tdiag.center-d1.resid-agree-stable\endcsname{0.000}
\expandafter\def\csname cnt@rev.desc.d6.uiv.p2.C.tdiag.center-d1.resid-agree-stable\endcsname{0}
\expandafter\def\csname pct@rev.desc.d6.uiv.p2.C.tdiag.center-d1.resid-agree-stable\endcsname{0.0}
\expandafter\def\csname ppt@rev.desc.d6.uiv.p2.C.tdiag.center-d1.resid-agree-stable\endcsname{0.0}
\expandafter\def\csname val@rev.desc.d6.uiv.p2.C.tdiag.center-d1.resid-disagree\endcsname{0.125}
\expandafter\def\csname cnt@rev.desc.d6.uiv.p2.C.tdiag.center-d1.resid-disagree\endcsname{0}
\expandafter\def\csname pct@rev.desc.d6.uiv.p2.C.tdiag.center-d1.resid-disagree\endcsname{12.5}
\expandafter\def\csname ppt@rev.desc.d6.uiv.p2.C.tdiag.center-d1.resid-disagree\endcsname{$+$12.5}
\expandafter\def\csname val@rev.desc.d6.uiv.p2.C.tdiag.center-d1.share-agree-stable\endcsname{0.906}
\expandafter\def\csname cnt@rev.desc.d6.uiv.p2.C.tdiag.center-d1.share-agree-stable\endcsname{1}
\expandafter\def\csname pct@rev.desc.d6.uiv.p2.C.tdiag.center-d1.share-agree-stable\endcsname{90.6}
\expandafter\def\csname ppt@rev.desc.d6.uiv.p2.C.tdiag.center-d1.share-agree-stable\endcsname{$+$90.6}
\expandafter\def\csname val@rev.desc.d6.uiv.p2.C.tdiag.context-d1.N-agree-stable\endcsname{83}
\expandafter\def\csname cnt@rev.desc.d6.uiv.p2.C.tdiag.context-d1.N-agree-stable\endcsname{83}
\expandafter\def\csname pct@rev.desc.d6.uiv.p2.C.tdiag.context-d1.N-agree-stable\endcsname{8300.0}
\expandafter\def\csname ppt@rev.desc.d6.uiv.p2.C.tdiag.context-d1.N-agree-stable\endcsname{$+$8300.0}
\expandafter\def\csname val@rev.desc.d6.uiv.p2.C.tdiag.context-d1.N-disagree\endcsname{11}
\expandafter\def\csname cnt@rev.desc.d6.uiv.p2.C.tdiag.context-d1.N-disagree\endcsname{11}
\expandafter\def\csname pct@rev.desc.d6.uiv.p2.C.tdiag.context-d1.N-disagree\endcsname{1100.0}
\expandafter\def\csname ppt@rev.desc.d6.uiv.p2.C.tdiag.context-d1.N-disagree\endcsname{$+$1100.0}
\expandafter\def\csname val@rev.desc.d6.uiv.p2.C.tdiag.context-d1.boot-resid.ci95.0\endcsname{-0.035}
\expandafter\def\csname cnt@rev.desc.d6.uiv.p2.C.tdiag.context-d1.boot-resid.ci95.0\endcsname{0}
\expandafter\def\csname pct@rev.desc.d6.uiv.p2.C.tdiag.context-d1.boot-resid.ci95.0\endcsname{$-$3.5}
\expandafter\def\csname ppt@rev.desc.d6.uiv.p2.C.tdiag.context-d1.boot-resid.ci95.0\endcsname{$-$3.5}
\expandafter\def\csname val@rev.desc.d6.uiv.p2.C.tdiag.context-d1.boot-resid.ci95.1\endcsname{0.037}
\expandafter\def\csname cnt@rev.desc.d6.uiv.p2.C.tdiag.context-d1.boot-resid.ci95.1\endcsname{0}
\expandafter\def\csname pct@rev.desc.d6.uiv.p2.C.tdiag.context-d1.boot-resid.ci95.1\endcsname{3.8}
\expandafter\def\csname ppt@rev.desc.d6.uiv.p2.C.tdiag.context-d1.boot-resid.ci95.1\endcsname{$+$3.8}
\expandafter\def\csname val@rev.desc.d6.uiv.p2.C.tdiag.context-d1.boot-resid.n-apps\endcsname{12}
\expandafter\def\csname cnt@rev.desc.d6.uiv.p2.C.tdiag.context-d1.boot-resid.n-apps\endcsname{12}
\expandafter\def\csname pct@rev.desc.d6.uiv.p2.C.tdiag.context-d1.boot-resid.n-apps\endcsname{1200.0}
\expandafter\def\csname ppt@rev.desc.d6.uiv.p2.C.tdiag.context-d1.boot-resid.n-apps\endcsname{$+$1200.0}
\expandafter\def\csname val@rev.desc.d6.uiv.p2.C.tdiag.context-d1.boot-resid.n-boot-valid\endcsname{4000}
\expandafter\def\csname cnt@rev.desc.d6.uiv.p2.C.tdiag.context-d1.boot-resid.n-boot-valid\endcsname{4{,}000}
\expandafter\def\csname pct@rev.desc.d6.uiv.p2.C.tdiag.context-d1.boot-resid.n-boot-valid\endcsname{400000.0}
\expandafter\def\csname ppt@rev.desc.d6.uiv.p2.C.tdiag.context-d1.boot-resid.n-boot-valid\endcsname{$+$400000.0}
\expandafter\def\csname val@rev.desc.d6.uiv.p2.C.tdiag.context-d1.boot-resid.value\endcsname{0.000}
\expandafter\def\csname cnt@rev.desc.d6.uiv.p2.C.tdiag.context-d1.boot-resid.value\endcsname{0}
\expandafter\def\csname pct@rev.desc.d6.uiv.p2.C.tdiag.context-d1.boot-resid.value\endcsname{0.0}
\expandafter\def\csname ppt@rev.desc.d6.uiv.p2.C.tdiag.context-d1.boot-resid.value\endcsname{0.0}
\expandafter\def\csname val@rev.desc.d6.uiv.p2.C.tdiag.context-d1.boot-share.ci95.0\endcsname{0.802}
\expandafter\def\csname cnt@rev.desc.d6.uiv.p2.C.tdiag.context-d1.boot-share.ci95.0\endcsname{1}
\expandafter\def\csname pct@rev.desc.d6.uiv.p2.C.tdiag.context-d1.boot-share.ci95.0\endcsname{80.2}
\expandafter\def\csname ppt@rev.desc.d6.uiv.p2.C.tdiag.context-d1.boot-share.ci95.0\endcsname{$+$80.2}
\expandafter\def\csname val@rev.desc.d6.uiv.p2.C.tdiag.context-d1.boot-share.ci95.1\endcsname{0.927}
\expandafter\def\csname cnt@rev.desc.d6.uiv.p2.C.tdiag.context-d1.boot-share.ci95.1\endcsname{1}
\expandafter\def\csname pct@rev.desc.d6.uiv.p2.C.tdiag.context-d1.boot-share.ci95.1\endcsname{92.7}
\expandafter\def\csname ppt@rev.desc.d6.uiv.p2.C.tdiag.context-d1.boot-share.ci95.1\endcsname{$+$92.7}
\expandafter\def\csname val@rev.desc.d6.uiv.p2.C.tdiag.context-d1.boot-share.n-apps\endcsname{12}
\expandafter\def\csname cnt@rev.desc.d6.uiv.p2.C.tdiag.context-d1.boot-share.n-apps\endcsname{12}
\expandafter\def\csname pct@rev.desc.d6.uiv.p2.C.tdiag.context-d1.boot-share.n-apps\endcsname{1200.0}
\expandafter\def\csname ppt@rev.desc.d6.uiv.p2.C.tdiag.context-d1.boot-share.n-apps\endcsname{$+$1200.0}
\expandafter\def\csname val@rev.desc.d6.uiv.p2.C.tdiag.context-d1.boot-share.n-boot-valid\endcsname{4000}
\expandafter\def\csname cnt@rev.desc.d6.uiv.p2.C.tdiag.context-d1.boot-share.n-boot-valid\endcsname{4{,}000}
\expandafter\def\csname pct@rev.desc.d6.uiv.p2.C.tdiag.context-d1.boot-share.n-boot-valid\endcsname{400000.0}
\expandafter\def\csname ppt@rev.desc.d6.uiv.p2.C.tdiag.context-d1.boot-share.n-boot-valid\endcsname{$+$400000.0}
\expandafter\def\csname val@rev.desc.d6.uiv.p2.C.tdiag.context-d1.boot-share.value\endcsname{0.865}
\expandafter\def\csname cnt@rev.desc.d6.uiv.p2.C.tdiag.context-d1.boot-share.value\endcsname{1}
\expandafter\def\csname pct@rev.desc.d6.uiv.p2.C.tdiag.context-d1.boot-share.value\endcsname{86.5}
\expandafter\def\csname ppt@rev.desc.d6.uiv.p2.C.tdiag.context-d1.boot-share.value\endcsname{$+$86.5}
\expandafter\def\csname val@rev.desc.d6.uiv.p2.C.tdiag.context-d1.n\endcsname{96}
\expandafter\def\csname cnt@rev.desc.d6.uiv.p2.C.tdiag.context-d1.n\endcsname{96}
\expandafter\def\csname pct@rev.desc.d6.uiv.p2.C.tdiag.context-d1.n\endcsname{9600.0}
\expandafter\def\csname ppt@rev.desc.d6.uiv.p2.C.tdiag.context-d1.n\endcsname{$+$9600.0}
\expandafter\def\csname val@rev.desc.d6.uiv.p2.C.tdiag.context-d1.resid-agree-stable\endcsname{0.000}
\expandafter\def\csname cnt@rev.desc.d6.uiv.p2.C.tdiag.context-d1.resid-agree-stable\endcsname{0}
\expandafter\def\csname pct@rev.desc.d6.uiv.p2.C.tdiag.context-d1.resid-agree-stable\endcsname{0.0}
\expandafter\def\csname ppt@rev.desc.d6.uiv.p2.C.tdiag.context-d1.resid-agree-stable\endcsname{0.0}
\expandafter\def\csname val@rev.desc.d6.uiv.p2.C.tdiag.context-d1.resid-disagree\endcsname{-0.091}
\expandafter\def\csname cnt@rev.desc.d6.uiv.p2.C.tdiag.context-d1.resid-disagree\endcsname{0}
\expandafter\def\csname pct@rev.desc.d6.uiv.p2.C.tdiag.context-d1.resid-disagree\endcsname{$-$9.1}
\expandafter\def\csname ppt@rev.desc.d6.uiv.p2.C.tdiag.context-d1.resid-disagree\endcsname{$-$9.1}
\expandafter\def\csname val@rev.desc.d6.uiv.p2.C.tdiag.context-d1.share-agree-stable\endcsname{0.865}
\expandafter\def\csname cnt@rev.desc.d6.uiv.p2.C.tdiag.context-d1.share-agree-stable\endcsname{1}
\expandafter\def\csname pct@rev.desc.d6.uiv.p2.C.tdiag.context-d1.share-agree-stable\endcsname{86.5}
\expandafter\def\csname ppt@rev.desc.d6.uiv.p2.C.tdiag.context-d1.share-agree-stable\endcsname{$+$86.5}
\expandafter\def\csname val@rev.desc.d6.uiv.p2.C.tdiag.dual-d2.N-agree-stable\endcsname{86}
\expandafter\def\csname cnt@rev.desc.d6.uiv.p2.C.tdiag.dual-d2.N-agree-stable\endcsname{86}
\expandafter\def\csname pct@rev.desc.d6.uiv.p2.C.tdiag.dual-d2.N-agree-stable\endcsname{8600.0}
\expandafter\def\csname ppt@rev.desc.d6.uiv.p2.C.tdiag.dual-d2.N-agree-stable\endcsname{$+$8600.0}
\expandafter\def\csname val@rev.desc.d6.uiv.p2.C.tdiag.dual-d2.N-disagree\endcsname{6}
\expandafter\def\csname cnt@rev.desc.d6.uiv.p2.C.tdiag.dual-d2.N-disagree\endcsname{6}
\expandafter\def\csname pct@rev.desc.d6.uiv.p2.C.tdiag.dual-d2.N-disagree\endcsname{600.0}
\expandafter\def\csname ppt@rev.desc.d6.uiv.p2.C.tdiag.dual-d2.N-disagree\endcsname{$+$600.0}
\expandafter\def\csname val@rev.desc.d6.uiv.p2.C.tdiag.dual-d2.boot-resid.ci95.0\endcsname{-0.043}
\expandafter\def\csname cnt@rev.desc.d6.uiv.p2.C.tdiag.dual-d2.boot-resid.ci95.0\endcsname{0}
\expandafter\def\csname pct@rev.desc.d6.uiv.p2.C.tdiag.dual-d2.boot-resid.ci95.0\endcsname{$-$4.3}
\expandafter\def\csname ppt@rev.desc.d6.uiv.p2.C.tdiag.dual-d2.boot-resid.ci95.0\endcsname{$-$4.3}
\expandafter\def\csname val@rev.desc.d6.uiv.p2.C.tdiag.dual-d2.boot-resid.ci95.1\endcsname{0.080}
\expandafter\def\csname cnt@rev.desc.d6.uiv.p2.C.tdiag.dual-d2.boot-resid.ci95.1\endcsname{0}
\expandafter\def\csname pct@rev.desc.d6.uiv.p2.C.tdiag.dual-d2.boot-resid.ci95.1\endcsname{8.0}
\expandafter\def\csname ppt@rev.desc.d6.uiv.p2.C.tdiag.dual-d2.boot-resid.ci95.1\endcsname{$+$8.0}
\expandafter\def\csname val@rev.desc.d6.uiv.p2.C.tdiag.dual-d2.boot-resid.n-apps\endcsname{12}
\expandafter\def\csname cnt@rev.desc.d6.uiv.p2.C.tdiag.dual-d2.boot-resid.n-apps\endcsname{12}
\expandafter\def\csname pct@rev.desc.d6.uiv.p2.C.tdiag.dual-d2.boot-resid.n-apps\endcsname{1200.0}
\expandafter\def\csname ppt@rev.desc.d6.uiv.p2.C.tdiag.dual-d2.boot-resid.n-apps\endcsname{$+$1200.0}
\expandafter\def\csname val@rev.desc.d6.uiv.p2.C.tdiag.dual-d2.boot-resid.n-boot-valid\endcsname{4000}
\expandafter\def\csname cnt@rev.desc.d6.uiv.p2.C.tdiag.dual-d2.boot-resid.n-boot-valid\endcsname{4{,}000}
\expandafter\def\csname pct@rev.desc.d6.uiv.p2.C.tdiag.dual-d2.boot-resid.n-boot-valid\endcsname{400000.0}
\expandafter\def\csname ppt@rev.desc.d6.uiv.p2.C.tdiag.dual-d2.boot-resid.n-boot-valid\endcsname{$+$400000.0}
\expandafter\def\csname val@rev.desc.d6.uiv.p2.C.tdiag.dual-d2.boot-resid.value\endcsname{0.012}
\expandafter\def\csname cnt@rev.desc.d6.uiv.p2.C.tdiag.dual-d2.boot-resid.value\endcsname{0}
\expandafter\def\csname pct@rev.desc.d6.uiv.p2.C.tdiag.dual-d2.boot-resid.value\endcsname{1.2}
\expandafter\def\csname ppt@rev.desc.d6.uiv.p2.C.tdiag.dual-d2.boot-resid.value\endcsname{$+$1.2}
\expandafter\def\csname val@rev.desc.d6.uiv.p2.C.tdiag.dual-d2.boot-share.ci95.0\endcsname{0.833}
\expandafter\def\csname cnt@rev.desc.d6.uiv.p2.C.tdiag.dual-d2.boot-share.ci95.0\endcsname{1}
\expandafter\def\csname pct@rev.desc.d6.uiv.p2.C.tdiag.dual-d2.boot-share.ci95.0\endcsname{83.3}
\expandafter\def\csname ppt@rev.desc.d6.uiv.p2.C.tdiag.dual-d2.boot-share.ci95.0\endcsname{$+$83.3}
\expandafter\def\csname val@rev.desc.d6.uiv.p2.C.tdiag.dual-d2.boot-share.ci95.1\endcsname{0.958}
\expandafter\def\csname cnt@rev.desc.d6.uiv.p2.C.tdiag.dual-d2.boot-share.ci95.1\endcsname{1}
\expandafter\def\csname pct@rev.desc.d6.uiv.p2.C.tdiag.dual-d2.boot-share.ci95.1\endcsname{95.8}
\expandafter\def\csname ppt@rev.desc.d6.uiv.p2.C.tdiag.dual-d2.boot-share.ci95.1\endcsname{$+$95.8}
\expandafter\def\csname val@rev.desc.d6.uiv.p2.C.tdiag.dual-d2.boot-share.n-apps\endcsname{12}
\expandafter\def\csname cnt@rev.desc.d6.uiv.p2.C.tdiag.dual-d2.boot-share.n-apps\endcsname{12}
\expandafter\def\csname pct@rev.desc.d6.uiv.p2.C.tdiag.dual-d2.boot-share.n-apps\endcsname{1200.0}
\expandafter\def\csname ppt@rev.desc.d6.uiv.p2.C.tdiag.dual-d2.boot-share.n-apps\endcsname{$+$1200.0}
\expandafter\def\csname val@rev.desc.d6.uiv.p2.C.tdiag.dual-d2.boot-share.n-boot-valid\endcsname{4000}
\expandafter\def\csname cnt@rev.desc.d6.uiv.p2.C.tdiag.dual-d2.boot-share.n-boot-valid\endcsname{4{,}000}
\expandafter\def\csname pct@rev.desc.d6.uiv.p2.C.tdiag.dual-d2.boot-share.n-boot-valid\endcsname{400000.0}
\expandafter\def\csname ppt@rev.desc.d6.uiv.p2.C.tdiag.dual-d2.boot-share.n-boot-valid\endcsname{$+$400000.0}
\expandafter\def\csname val@rev.desc.d6.uiv.p2.C.tdiag.dual-d2.boot-share.value\endcsname{0.896}
\expandafter\def\csname cnt@rev.desc.d6.uiv.p2.C.tdiag.dual-d2.boot-share.value\endcsname{1}
\expandafter\def\csname pct@rev.desc.d6.uiv.p2.C.tdiag.dual-d2.boot-share.value\endcsname{89.6}
\expandafter\def\csname ppt@rev.desc.d6.uiv.p2.C.tdiag.dual-d2.boot-share.value\endcsname{$+$89.6}
\expandafter\def\csname val@rev.desc.d6.uiv.p2.C.tdiag.dual-d2.n\endcsname{96}
\expandafter\def\csname cnt@rev.desc.d6.uiv.p2.C.tdiag.dual-d2.n\endcsname{96}
\expandafter\def\csname pct@rev.desc.d6.uiv.p2.C.tdiag.dual-d2.n\endcsname{9600.0}
\expandafter\def\csname ppt@rev.desc.d6.uiv.p2.C.tdiag.dual-d2.n\endcsname{$+$9600.0}
\expandafter\def\csname val@rev.desc.d6.uiv.p2.C.tdiag.dual-d2.resid-agree-stable\endcsname{0.012}
\expandafter\def\csname cnt@rev.desc.d6.uiv.p2.C.tdiag.dual-d2.resid-agree-stable\endcsname{0}
\expandafter\def\csname pct@rev.desc.d6.uiv.p2.C.tdiag.dual-d2.resid-agree-stable\endcsname{1.2}
\expandafter\def\csname ppt@rev.desc.d6.uiv.p2.C.tdiag.dual-d2.resid-agree-stable\endcsname{$+$1.2}
\expandafter\def\csname val@rev.desc.d6.uiv.p2.C.tdiag.dual-d2.resid-disagree\endcsname{-0.333}
\expandafter\def\csname cnt@rev.desc.d6.uiv.p2.C.tdiag.dual-d2.resid-disagree\endcsname{0}
\expandafter\def\csname pct@rev.desc.d6.uiv.p2.C.tdiag.dual-d2.resid-disagree\endcsname{$-$33.3}
\expandafter\def\csname ppt@rev.desc.d6.uiv.p2.C.tdiag.dual-d2.resid-disagree\endcsname{$-$33.3}
\expandafter\def\csname val@rev.desc.d6.uiv.p2.C.tdiag.dual-d2.share-agree-stable\endcsname{0.896}
\expandafter\def\csname cnt@rev.desc.d6.uiv.p2.C.tdiag.dual-d2.share-agree-stable\endcsname{1}
\expandafter\def\csname pct@rev.desc.d6.uiv.p2.C.tdiag.dual-d2.share-agree-stable\endcsname{89.6}
\expandafter\def\csname ppt@rev.desc.d6.uiv.p2.C.tdiag.dual-d2.share-agree-stable\endcsname{$+$89.6}
\expandafter\def\csname val@rev.desc.d6.uiv.p2.C.tdiag.grid2-d3.N-agree-stable\endcsname{78}
\expandafter\def\csname cnt@rev.desc.d6.uiv.p2.C.tdiag.grid2-d3.N-agree-stable\endcsname{78}
\expandafter\def\csname pct@rev.desc.d6.uiv.p2.C.tdiag.grid2-d3.N-agree-stable\endcsname{7800.0}
\expandafter\def\csname ppt@rev.desc.d6.uiv.p2.C.tdiag.grid2-d3.N-agree-stable\endcsname{$+$7800.0}
\expandafter\def\csname val@rev.desc.d6.uiv.p2.C.tdiag.grid2-d3.N-disagree\endcsname{15}
\expandafter\def\csname cnt@rev.desc.d6.uiv.p2.C.tdiag.grid2-d3.N-disagree\endcsname{15}
\expandafter\def\csname pct@rev.desc.d6.uiv.p2.C.tdiag.grid2-d3.N-disagree\endcsname{1500.0}
\expandafter\def\csname ppt@rev.desc.d6.uiv.p2.C.tdiag.grid2-d3.N-disagree\endcsname{$+$1500.0}
\expandafter\def\csname val@rev.desc.d6.uiv.p2.C.tdiag.grid2-d3.boot-resid.ci95.0\endcsname{-0.025}
\expandafter\def\csname cnt@rev.desc.d6.uiv.p2.C.tdiag.grid2-d3.boot-resid.ci95.0\endcsname{0}
\expandafter\def\csname pct@rev.desc.d6.uiv.p2.C.tdiag.grid2-d3.boot-resid.ci95.0\endcsname{$-$2.5}
\expandafter\def\csname ppt@rev.desc.d6.uiv.p2.C.tdiag.grid2-d3.boot-resid.ci95.0\endcsname{$-$2.5}
\expandafter\def\csname val@rev.desc.d6.uiv.p2.C.tdiag.grid2-d3.boot-resid.ci95.1\endcsname{0.075}
\expandafter\def\csname cnt@rev.desc.d6.uiv.p2.C.tdiag.grid2-d3.boot-resid.ci95.1\endcsname{0}
\expandafter\def\csname pct@rev.desc.d6.uiv.p2.C.tdiag.grid2-d3.boot-resid.ci95.1\endcsname{7.5}
\expandafter\def\csname ppt@rev.desc.d6.uiv.p2.C.tdiag.grid2-d3.boot-resid.ci95.1\endcsname{$+$7.5}
\expandafter\def\csname val@rev.desc.d6.uiv.p2.C.tdiag.grid2-d3.boot-resid.n-apps\endcsname{12}
\expandafter\def\csname cnt@rev.desc.d6.uiv.p2.C.tdiag.grid2-d3.boot-resid.n-apps\endcsname{12}
\expandafter\def\csname pct@rev.desc.d6.uiv.p2.C.tdiag.grid2-d3.boot-resid.n-apps\endcsname{1200.0}
\expandafter\def\csname ppt@rev.desc.d6.uiv.p2.C.tdiag.grid2-d3.boot-resid.n-apps\endcsname{$+$1200.0}
\expandafter\def\csname val@rev.desc.d6.uiv.p2.C.tdiag.grid2-d3.boot-resid.n-boot-valid\endcsname{4000}
\expandafter\def\csname cnt@rev.desc.d6.uiv.p2.C.tdiag.grid2-d3.boot-resid.n-boot-valid\endcsname{4{,}000}
\expandafter\def\csname pct@rev.desc.d6.uiv.p2.C.tdiag.grid2-d3.boot-resid.n-boot-valid\endcsname{400000.0}
\expandafter\def\csname ppt@rev.desc.d6.uiv.p2.C.tdiag.grid2-d3.boot-resid.n-boot-valid\endcsname{$+$400000.0}
\expandafter\def\csname val@rev.desc.d6.uiv.p2.C.tdiag.grid2-d3.boot-resid.value\endcsname{0.026}
\expandafter\def\csname cnt@rev.desc.d6.uiv.p2.C.tdiag.grid2-d3.boot-resid.value\endcsname{0}
\expandafter\def\csname pct@rev.desc.d6.uiv.p2.C.tdiag.grid2-d3.boot-resid.value\endcsname{2.6}
\expandafter\def\csname ppt@rev.desc.d6.uiv.p2.C.tdiag.grid2-d3.boot-resid.value\endcsname{$+$2.6}
\expandafter\def\csname val@rev.desc.d6.uiv.p2.C.tdiag.grid2-d3.boot-share.ci95.0\endcsname{0.740}
\expandafter\def\csname cnt@rev.desc.d6.uiv.p2.C.tdiag.grid2-d3.boot-share.ci95.0\endcsname{1}
\expandafter\def\csname pct@rev.desc.d6.uiv.p2.C.tdiag.grid2-d3.boot-share.ci95.0\endcsname{74.0}
\expandafter\def\csname ppt@rev.desc.d6.uiv.p2.C.tdiag.grid2-d3.boot-share.ci95.0\endcsname{$+$74.0}
\expandafter\def\csname val@rev.desc.d6.uiv.p2.C.tdiag.grid2-d3.boot-share.ci95.1\endcsname{0.885}
\expandafter\def\csname cnt@rev.desc.d6.uiv.p2.C.tdiag.grid2-d3.boot-share.ci95.1\endcsname{1}
\expandafter\def\csname pct@rev.desc.d6.uiv.p2.C.tdiag.grid2-d3.boot-share.ci95.1\endcsname{88.5}
\expandafter\def\csname ppt@rev.desc.d6.uiv.p2.C.tdiag.grid2-d3.boot-share.ci95.1\endcsname{$+$88.5}
\expandafter\def\csname val@rev.desc.d6.uiv.p2.C.tdiag.grid2-d3.boot-share.n-apps\endcsname{12}
\expandafter\def\csname cnt@rev.desc.d6.uiv.p2.C.tdiag.grid2-d3.boot-share.n-apps\endcsname{12}
\expandafter\def\csname pct@rev.desc.d6.uiv.p2.C.tdiag.grid2-d3.boot-share.n-apps\endcsname{1200.0}
\expandafter\def\csname ppt@rev.desc.d6.uiv.p2.C.tdiag.grid2-d3.boot-share.n-apps\endcsname{$+$1200.0}
\expandafter\def\csname val@rev.desc.d6.uiv.p2.C.tdiag.grid2-d3.boot-share.n-boot-valid\endcsname{4000}
\expandafter\def\csname cnt@rev.desc.d6.uiv.p2.C.tdiag.grid2-d3.boot-share.n-boot-valid\endcsname{4{,}000}
\expandafter\def\csname pct@rev.desc.d6.uiv.p2.C.tdiag.grid2-d3.boot-share.n-boot-valid\endcsname{400000.0}
\expandafter\def\csname ppt@rev.desc.d6.uiv.p2.C.tdiag.grid2-d3.boot-share.n-boot-valid\endcsname{$+$400000.0}
\expandafter\def\csname val@rev.desc.d6.uiv.p2.C.tdiag.grid2-d3.boot-share.value\endcsname{0.812}
\expandafter\def\csname cnt@rev.desc.d6.uiv.p2.C.tdiag.grid2-d3.boot-share.value\endcsname{1}
\expandafter\def\csname pct@rev.desc.d6.uiv.p2.C.tdiag.grid2-d3.boot-share.value\endcsname{81.2}
\expandafter\def\csname ppt@rev.desc.d6.uiv.p2.C.tdiag.grid2-d3.boot-share.value\endcsname{$+$81.2}
\expandafter\def\csname val@rev.desc.d6.uiv.p2.C.tdiag.grid2-d3.n\endcsname{96}
\expandafter\def\csname cnt@rev.desc.d6.uiv.p2.C.tdiag.grid2-d3.n\endcsname{96}
\expandafter\def\csname pct@rev.desc.d6.uiv.p2.C.tdiag.grid2-d3.n\endcsname{9600.0}
\expandafter\def\csname ppt@rev.desc.d6.uiv.p2.C.tdiag.grid2-d3.n\endcsname{$+$9600.0}
\expandafter\def\csname val@rev.desc.d6.uiv.p2.C.tdiag.grid2-d3.resid-agree-stable\endcsname{0.026}
\expandafter\def\csname cnt@rev.desc.d6.uiv.p2.C.tdiag.grid2-d3.resid-agree-stable\endcsname{0}
\expandafter\def\csname pct@rev.desc.d6.uiv.p2.C.tdiag.grid2-d3.resid-agree-stable\endcsname{2.6}
\expandafter\def\csname ppt@rev.desc.d6.uiv.p2.C.tdiag.grid2-d3.resid-agree-stable\endcsname{$+$2.6}
\expandafter\def\csname val@rev.desc.d6.uiv.p2.C.tdiag.grid2-d3.resid-disagree\endcsname{0.200}
\expandafter\def\csname cnt@rev.desc.d6.uiv.p2.C.tdiag.grid2-d3.resid-disagree\endcsname{0}
\expandafter\def\csname pct@rev.desc.d6.uiv.p2.C.tdiag.grid2-d3.resid-disagree\endcsname{20.0}
\expandafter\def\csname ppt@rev.desc.d6.uiv.p2.C.tdiag.grid2-d3.resid-disagree\endcsname{$+$20.0}
\expandafter\def\csname val@rev.desc.d6.uiv.p2.C.tdiag.grid2-d3.share-agree-stable\endcsname{0.812}
\expandafter\def\csname cnt@rev.desc.d6.uiv.p2.C.tdiag.grid2-d3.share-agree-stable\endcsname{1}
\expandafter\def\csname pct@rev.desc.d6.uiv.p2.C.tdiag.grid2-d3.share-agree-stable\endcsname{81.2}
\expandafter\def\csname ppt@rev.desc.d6.uiv.p2.C.tdiag.grid2-d3.share-agree-stable\endcsname{$+$81.2}
\expandafter\def\csname val@rev.desc.d6.uiv.p2.C.tdiag.zoomclick-d3.N-agree-stable\endcsname{84}
\expandafter\def\csname cnt@rev.desc.d6.uiv.p2.C.tdiag.zoomclick-d3.N-agree-stable\endcsname{84}
\expandafter\def\csname pct@rev.desc.d6.uiv.p2.C.tdiag.zoomclick-d3.N-agree-stable\endcsname{8400.0}
\expandafter\def\csname ppt@rev.desc.d6.uiv.p2.C.tdiag.zoomclick-d3.N-agree-stable\endcsname{$+$8400.0}
\expandafter\def\csname val@rev.desc.d6.uiv.p2.C.tdiag.zoomclick-d3.N-disagree\endcsname{12}
\expandafter\def\csname cnt@rev.desc.d6.uiv.p2.C.tdiag.zoomclick-d3.N-disagree\endcsname{12}
\expandafter\def\csname pct@rev.desc.d6.uiv.p2.C.tdiag.zoomclick-d3.N-disagree\endcsname{1200.0}
\expandafter\def\csname ppt@rev.desc.d6.uiv.p2.C.tdiag.zoomclick-d3.N-disagree\endcsname{$+$1200.0}
\expandafter\def\csname val@rev.desc.d6.uiv.p2.C.tdiag.zoomclick-d3.boot-resid.ci95.0\endcsname{-0.023}
\expandafter\def\csname cnt@rev.desc.d6.uiv.p2.C.tdiag.zoomclick-d3.boot-resid.ci95.0\endcsname{0}
\expandafter\def\csname pct@rev.desc.d6.uiv.p2.C.tdiag.zoomclick-d3.boot-resid.ci95.0\endcsname{$-$2.3}
\expandafter\def\csname ppt@rev.desc.d6.uiv.p2.C.tdiag.zoomclick-d3.boot-resid.ci95.0\endcsname{$-$2.3}
\expandafter\def\csname val@rev.desc.d6.uiv.p2.C.tdiag.zoomclick-d3.boot-resid.ci95.1\endcsname{0.071}
\expandafter\def\csname cnt@rev.desc.d6.uiv.p2.C.tdiag.zoomclick-d3.boot-resid.ci95.1\endcsname{0}
\expandafter\def\csname pct@rev.desc.d6.uiv.p2.C.tdiag.zoomclick-d3.boot-resid.ci95.1\endcsname{7.1}
\expandafter\def\csname ppt@rev.desc.d6.uiv.p2.C.tdiag.zoomclick-d3.boot-resid.ci95.1\endcsname{$+$7.1}
\expandafter\def\csname val@rev.desc.d6.uiv.p2.C.tdiag.zoomclick-d3.boot-resid.n-apps\endcsname{12}
\expandafter\def\csname cnt@rev.desc.d6.uiv.p2.C.tdiag.zoomclick-d3.boot-resid.n-apps\endcsname{12}
\expandafter\def\csname pct@rev.desc.d6.uiv.p2.C.tdiag.zoomclick-d3.boot-resid.n-apps\endcsname{1200.0}
\expandafter\def\csname ppt@rev.desc.d6.uiv.p2.C.tdiag.zoomclick-d3.boot-resid.n-apps\endcsname{$+$1200.0}
\expandafter\def\csname val@rev.desc.d6.uiv.p2.C.tdiag.zoomclick-d3.boot-resid.n-boot-valid\endcsname{4000}
\expandafter\def\csname cnt@rev.desc.d6.uiv.p2.C.tdiag.zoomclick-d3.boot-resid.n-boot-valid\endcsname{4{,}000}
\expandafter\def\csname pct@rev.desc.d6.uiv.p2.C.tdiag.zoomclick-d3.boot-resid.n-boot-valid\endcsname{400000.0}
\expandafter\def\csname ppt@rev.desc.d6.uiv.p2.C.tdiag.zoomclick-d3.boot-resid.n-boot-valid\endcsname{$+$400000.0}
\expandafter\def\csname val@rev.desc.d6.uiv.p2.C.tdiag.zoomclick-d3.boot-resid.value\endcsname{0.024}
\expandafter\def\csname cnt@rev.desc.d6.uiv.p2.C.tdiag.zoomclick-d3.boot-resid.value\endcsname{0}
\expandafter\def\csname pct@rev.desc.d6.uiv.p2.C.tdiag.zoomclick-d3.boot-resid.value\endcsname{2.4}
\expandafter\def\csname ppt@rev.desc.d6.uiv.p2.C.tdiag.zoomclick-d3.boot-resid.value\endcsname{$+$2.4}
\expandafter\def\csname val@rev.desc.d6.uiv.p2.C.tdiag.zoomclick-d3.boot-share.ci95.0\endcsname{0.802}
\expandafter\def\csname cnt@rev.desc.d6.uiv.p2.C.tdiag.zoomclick-d3.boot-share.ci95.0\endcsname{1}
\expandafter\def\csname pct@rev.desc.d6.uiv.p2.C.tdiag.zoomclick-d3.boot-share.ci95.0\endcsname{80.2}
\expandafter\def\csname ppt@rev.desc.d6.uiv.p2.C.tdiag.zoomclick-d3.boot-share.ci95.0\endcsname{$+$80.2}
\expandafter\def\csname val@rev.desc.d6.uiv.p2.C.tdiag.zoomclick-d3.boot-share.ci95.1\endcsname{0.938}
\expandafter\def\csname cnt@rev.desc.d6.uiv.p2.C.tdiag.zoomclick-d3.boot-share.ci95.1\endcsname{1}
\expandafter\def\csname pct@rev.desc.d6.uiv.p2.C.tdiag.zoomclick-d3.boot-share.ci95.1\endcsname{93.8}
\expandafter\def\csname ppt@rev.desc.d6.uiv.p2.C.tdiag.zoomclick-d3.boot-share.ci95.1\endcsname{$+$93.8}
\expandafter\def\csname val@rev.desc.d6.uiv.p2.C.tdiag.zoomclick-d3.boot-share.n-apps\endcsname{12}
\expandafter\def\csname cnt@rev.desc.d6.uiv.p2.C.tdiag.zoomclick-d3.boot-share.n-apps\endcsname{12}
\expandafter\def\csname pct@rev.desc.d6.uiv.p2.C.tdiag.zoomclick-d3.boot-share.n-apps\endcsname{1200.0}
\expandafter\def\csname ppt@rev.desc.d6.uiv.p2.C.tdiag.zoomclick-d3.boot-share.n-apps\endcsname{$+$1200.0}
\expandafter\def\csname val@rev.desc.d6.uiv.p2.C.tdiag.zoomclick-d3.boot-share.n-boot-valid\endcsname{4000}
\expandafter\def\csname cnt@rev.desc.d6.uiv.p2.C.tdiag.zoomclick-d3.boot-share.n-boot-valid\endcsname{4{,}000}
\expandafter\def\csname pct@rev.desc.d6.uiv.p2.C.tdiag.zoomclick-d3.boot-share.n-boot-valid\endcsname{400000.0}
\expandafter\def\csname ppt@rev.desc.d6.uiv.p2.C.tdiag.zoomclick-d3.boot-share.n-boot-valid\endcsname{$+$400000.0}
\expandafter\def\csname val@rev.desc.d6.uiv.p2.C.tdiag.zoomclick-d3.boot-share.value\endcsname{0.875}
\expandafter\def\csname cnt@rev.desc.d6.uiv.p2.C.tdiag.zoomclick-d3.boot-share.value\endcsname{1}
\expandafter\def\csname pct@rev.desc.d6.uiv.p2.C.tdiag.zoomclick-d3.boot-share.value\endcsname{87.5}
\expandafter\def\csname ppt@rev.desc.d6.uiv.p2.C.tdiag.zoomclick-d3.boot-share.value\endcsname{$+$87.5}
\expandafter\def\csname val@rev.desc.d6.uiv.p2.C.tdiag.zoomclick-d3.n\endcsname{96}
\expandafter\def\csname cnt@rev.desc.d6.uiv.p2.C.tdiag.zoomclick-d3.n\endcsname{96}
\expandafter\def\csname pct@rev.desc.d6.uiv.p2.C.tdiag.zoomclick-d3.n\endcsname{9600.0}
\expandafter\def\csname ppt@rev.desc.d6.uiv.p2.C.tdiag.zoomclick-d3.n\endcsname{$+$9600.0}
\expandafter\def\csname val@rev.desc.d6.uiv.p2.C.tdiag.zoomclick-d3.resid-agree-stable\endcsname{0.024}
\expandafter\def\csname cnt@rev.desc.d6.uiv.p2.C.tdiag.zoomclick-d3.resid-agree-stable\endcsname{0}
\expandafter\def\csname pct@rev.desc.d6.uiv.p2.C.tdiag.zoomclick-d3.resid-agree-stable\endcsname{2.4}
\expandafter\def\csname ppt@rev.desc.d6.uiv.p2.C.tdiag.zoomclick-d3.resid-agree-stable\endcsname{$+$2.4}
\expandafter\def\csname val@rev.desc.d6.uiv.p2.C.tdiag.zoomclick-d3.resid-disagree\endcsname{0.167}
\expandafter\def\csname cnt@rev.desc.d6.uiv.p2.C.tdiag.zoomclick-d3.resid-disagree\endcsname{0}
\expandafter\def\csname pct@rev.desc.d6.uiv.p2.C.tdiag.zoomclick-d3.resid-disagree\endcsname{16.7}
\expandafter\def\csname ppt@rev.desc.d6.uiv.p2.C.tdiag.zoomclick-d3.resid-disagree\endcsname{$+$16.7}
\expandafter\def\csname val@rev.desc.d6.uiv.p2.C.tdiag.zoomclick-d3.share-agree-stable\endcsname{0.875}
\expandafter\def\csname cnt@rev.desc.d6.uiv.p2.C.tdiag.zoomclick-d3.share-agree-stable\endcsname{1}
\expandafter\def\csname pct@rev.desc.d6.uiv.p2.C.tdiag.zoomclick-d3.share-agree-stable\endcsname{87.5}
\expandafter\def\csname ppt@rev.desc.d6.uiv.p2.C.tdiag.zoomclick-d3.share-agree-stable\endcsname{$+$87.5}
\expandafter\def\csname val@rev.desc.d6.uiv.p2.T.box.catalog-pooled.N-agree-stable\endcsname{441}
\expandafter\def\csname cnt@rev.desc.d6.uiv.p2.T.box.catalog-pooled.N-agree-stable\endcsname{441}
\expandafter\def\csname pct@rev.desc.d6.uiv.p2.T.box.catalog-pooled.N-agree-stable\endcsname{44100.0}
\expandafter\def\csname ppt@rev.desc.d6.uiv.p2.T.box.catalog-pooled.N-agree-stable\endcsname{$+$44100.0}
\expandafter\def\csname val@rev.desc.d6.uiv.p2.T.box.catalog-pooled.N-disagree\endcsname{122}
\expandafter\def\csname cnt@rev.desc.d6.uiv.p2.T.box.catalog-pooled.N-disagree\endcsname{122}
\expandafter\def\csname pct@rev.desc.d6.uiv.p2.T.box.catalog-pooled.N-disagree\endcsname{12200.0}
\expandafter\def\csname ppt@rev.desc.d6.uiv.p2.T.box.catalog-pooled.N-disagree\endcsname{$+$12200.0}
\expandafter\def\csname val@rev.desc.d6.uiv.p2.T.box.catalog-pooled.boot-resid.ci95.0\endcsname{-0.041}
\expandafter\def\csname cnt@rev.desc.d6.uiv.p2.T.box.catalog-pooled.boot-resid.ci95.0\endcsname{0}
\expandafter\def\csname pct@rev.desc.d6.uiv.p2.T.box.catalog-pooled.boot-resid.ci95.0\endcsname{$-$4.1}
\expandafter\def\csname ppt@rev.desc.d6.uiv.p2.T.box.catalog-pooled.boot-resid.ci95.0\endcsname{$-$4.1}
\expandafter\def\csname val@rev.desc.d6.uiv.p2.T.box.catalog-pooled.boot-resid.ci95.1\endcsname{-0.007}
\expandafter\def\csname cnt@rev.desc.d6.uiv.p2.T.box.catalog-pooled.boot-resid.ci95.1\endcsname{0}
\expandafter\def\csname pct@rev.desc.d6.uiv.p2.T.box.catalog-pooled.boot-resid.ci95.1\endcsname{$-$0.7}
\expandafter\def\csname ppt@rev.desc.d6.uiv.p2.T.box.catalog-pooled.boot-resid.ci95.1\endcsname{$-$0.7}
\expandafter\def\csname val@rev.desc.d6.uiv.p2.T.box.catalog-pooled.boot-resid.n-apps\endcsname{24}
\expandafter\def\csname cnt@rev.desc.d6.uiv.p2.T.box.catalog-pooled.boot-resid.n-apps\endcsname{24}
\expandafter\def\csname pct@rev.desc.d6.uiv.p2.T.box.catalog-pooled.boot-resid.n-apps\endcsname{2400.0}
\expandafter\def\csname ppt@rev.desc.d6.uiv.p2.T.box.catalog-pooled.boot-resid.n-apps\endcsname{$+$2400.0}
\expandafter\def\csname val@rev.desc.d6.uiv.p2.T.box.catalog-pooled.boot-resid.n-boot-valid\endcsname{4000}
\expandafter\def\csname cnt@rev.desc.d6.uiv.p2.T.box.catalog-pooled.boot-resid.n-boot-valid\endcsname{4{,}000}
\expandafter\def\csname pct@rev.desc.d6.uiv.p2.T.box.catalog-pooled.boot-resid.n-boot-valid\endcsname{400000.0}
\expandafter\def\csname ppt@rev.desc.d6.uiv.p2.T.box.catalog-pooled.boot-resid.n-boot-valid\endcsname{$+$400000.0}
\expandafter\def\csname val@rev.desc.d6.uiv.p2.T.box.catalog-pooled.boot-resid.value\endcsname{-0.023}
\expandafter\def\csname cnt@rev.desc.d6.uiv.p2.T.box.catalog-pooled.boot-resid.value\endcsname{0}
\expandafter\def\csname pct@rev.desc.d6.uiv.p2.T.box.catalog-pooled.boot-resid.value\endcsname{$-$2.3}
\expandafter\def\csname ppt@rev.desc.d6.uiv.p2.T.box.catalog-pooled.boot-resid.value\endcsname{$-$2.3}
\expandafter\def\csname val@rev.desc.d6.uiv.p2.T.box.catalog-pooled.boot-share.ci95.0\endcsname{0.700}
\expandafter\def\csname cnt@rev.desc.d6.uiv.p2.T.box.catalog-pooled.boot-share.ci95.0\endcsname{1}
\expandafter\def\csname pct@rev.desc.d6.uiv.p2.T.box.catalog-pooled.boot-share.ci95.0\endcsname{70.0}
\expandafter\def\csname ppt@rev.desc.d6.uiv.p2.T.box.catalog-pooled.boot-share.ci95.0\endcsname{$+$70.0}
\expandafter\def\csname val@rev.desc.d6.uiv.p2.T.box.catalog-pooled.boot-share.ci95.1\endcsname{0.828}
\expandafter\def\csname cnt@rev.desc.d6.uiv.p2.T.box.catalog-pooled.boot-share.ci95.1\endcsname{1}
\expandafter\def\csname pct@rev.desc.d6.uiv.p2.T.box.catalog-pooled.boot-share.ci95.1\endcsname{82.8}
\expandafter\def\csname ppt@rev.desc.d6.uiv.p2.T.box.catalog-pooled.boot-share.ci95.1\endcsname{$+$82.8}
\expandafter\def\csname val@rev.desc.d6.uiv.p2.T.box.catalog-pooled.boot-share.n-apps\endcsname{24}
\expandafter\def\csname cnt@rev.desc.d6.uiv.p2.T.box.catalog-pooled.boot-share.n-apps\endcsname{24}
\expandafter\def\csname pct@rev.desc.d6.uiv.p2.T.box.catalog-pooled.boot-share.n-apps\endcsname{2400.0}
\expandafter\def\csname ppt@rev.desc.d6.uiv.p2.T.box.catalog-pooled.boot-share.n-apps\endcsname{$+$2400.0}
\expandafter\def\csname val@rev.desc.d6.uiv.p2.T.box.catalog-pooled.boot-share.n-boot-valid\endcsname{4000}
\expandafter\def\csname cnt@rev.desc.d6.uiv.p2.T.box.catalog-pooled.boot-share.n-boot-valid\endcsname{4{,}000}
\expandafter\def\csname pct@rev.desc.d6.uiv.p2.T.box.catalog-pooled.boot-share.n-boot-valid\endcsname{400000.0}
\expandafter\def\csname ppt@rev.desc.d6.uiv.p2.T.box.catalog-pooled.boot-share.n-boot-valid\endcsname{$+$400000.0}
\expandafter\def\csname val@rev.desc.d6.uiv.p2.T.box.catalog-pooled.boot-share.value\endcsname{0.766}
\expandafter\def\csname cnt@rev.desc.d6.uiv.p2.T.box.catalog-pooled.boot-share.value\endcsname{1}
\expandafter\def\csname pct@rev.desc.d6.uiv.p2.T.box.catalog-pooled.boot-share.value\endcsname{76.6}
\expandafter\def\csname ppt@rev.desc.d6.uiv.p2.T.box.catalog-pooled.boot-share.value\endcsname{$+$76.6}
\expandafter\def\csname val@rev.desc.d6.uiv.p2.T.box.catalog-pooled.gap\endcsname{0.154}
\expandafter\def\csname cnt@rev.desc.d6.uiv.p2.T.box.catalog-pooled.gap\endcsname{0}
\expandafter\def\csname pct@rev.desc.d6.uiv.p2.T.box.catalog-pooled.gap\endcsname{15.4}
\expandafter\def\csname ppt@rev.desc.d6.uiv.p2.T.box.catalog-pooled.gap\endcsname{$+$15.4}
\expandafter\def\csname val@rev.desc.d6.uiv.p2.T.box.catalog-pooled.n\endcsname{576}
\expandafter\def\csname cnt@rev.desc.d6.uiv.p2.T.box.catalog-pooled.n\endcsname{576}
\expandafter\def\csname pct@rev.desc.d6.uiv.p2.T.box.catalog-pooled.n\endcsname{57600.0}
\expandafter\def\csname ppt@rev.desc.d6.uiv.p2.T.box.catalog-pooled.n\endcsname{$+$57600.0}
\expandafter\def\csname val@rev.desc.d6.uiv.p2.T.box.catalog-pooled.resid-agree-stable\endcsname{-0.023}
\expandafter\def\csname cnt@rev.desc.d6.uiv.p2.T.box.catalog-pooled.resid-agree-stable\endcsname{0}
\expandafter\def\csname pct@rev.desc.d6.uiv.p2.T.box.catalog-pooled.resid-agree-stable\endcsname{$-$2.3}
\expandafter\def\csname ppt@rev.desc.d6.uiv.p2.T.box.catalog-pooled.resid-agree-stable\endcsname{$-$2.3}
\expandafter\def\csname val@rev.desc.d6.uiv.p2.T.box.catalog-pooled.resid-disagree\endcsname{0.131}
\expandafter\def\csname cnt@rev.desc.d6.uiv.p2.T.box.catalog-pooled.resid-disagree\endcsname{0}
\expandafter\def\csname pct@rev.desc.d6.uiv.p2.T.box.catalog-pooled.resid-disagree\endcsname{13.1}
\expandafter\def\csname ppt@rev.desc.d6.uiv.p2.T.box.catalog-pooled.resid-disagree\endcsname{$+$13.1}
\expandafter\def\csname val@rev.desc.d6.uiv.p2.T.box.catalog-pooled.share-agree-stable\endcsname{0.766}
\expandafter\def\csname cnt@rev.desc.d6.uiv.p2.T.box.catalog-pooled.share-agree-stable\endcsname{1}
\expandafter\def\csname pct@rev.desc.d6.uiv.p2.T.box.catalog-pooled.share-agree-stable\endcsname{76.6}
\expandafter\def\csname ppt@rev.desc.d6.uiv.p2.T.box.catalog-pooled.share-agree-stable\endcsname{$+$76.6}
\expandafter\def\csname val@rev.desc.d6.uiv.p2.T.box.center-d1.N-agree-stable\endcsname{111}
\expandafter\def\csname cnt@rev.desc.d6.uiv.p2.T.box.center-d1.N-agree-stable\endcsname{111}
\expandafter\def\csname pct@rev.desc.d6.uiv.p2.T.box.center-d1.N-agree-stable\endcsname{11100.0}
\expandafter\def\csname ppt@rev.desc.d6.uiv.p2.T.box.center-d1.N-agree-stable\endcsname{$+$11100.0}
\expandafter\def\csname val@rev.desc.d6.uiv.p2.T.box.center-d1.N-disagree\endcsname{28}
\expandafter\def\csname cnt@rev.desc.d6.uiv.p2.T.box.center-d1.N-disagree\endcsname{28}
\expandafter\def\csname pct@rev.desc.d6.uiv.p2.T.box.center-d1.N-disagree\endcsname{2800.0}
\expandafter\def\csname ppt@rev.desc.d6.uiv.p2.T.box.center-d1.N-disagree\endcsname{$+$2800.0}
\expandafter\def\csname val@rev.desc.d6.uiv.p2.T.box.center-d1.boot-resid.ci95.0\endcsname{-0.028}
\expandafter\def\csname cnt@rev.desc.d6.uiv.p2.T.box.center-d1.boot-resid.ci95.0\endcsname{0}
\expandafter\def\csname pct@rev.desc.d6.uiv.p2.T.box.center-d1.boot-resid.ci95.0\endcsname{$-$2.8}
\expandafter\def\csname ppt@rev.desc.d6.uiv.p2.T.box.center-d1.boot-resid.ci95.0\endcsname{$-$2.8}
\expandafter\def\csname val@rev.desc.d6.uiv.p2.T.box.center-d1.boot-resid.ci95.1\endcsname{0.000}
\expandafter\def\csname cnt@rev.desc.d6.uiv.p2.T.box.center-d1.boot-resid.ci95.1\endcsname{0}
\expandafter\def\csname pct@rev.desc.d6.uiv.p2.T.box.center-d1.boot-resid.ci95.1\endcsname{0.0}
\expandafter\def\csname ppt@rev.desc.d6.uiv.p2.T.box.center-d1.boot-resid.ci95.1\endcsname{0.0}
\expandafter\def\csname val@rev.desc.d6.uiv.p2.T.box.center-d1.boot-resid.n-apps\endcsname{24}
\expandafter\def\csname cnt@rev.desc.d6.uiv.p2.T.box.center-d1.boot-resid.n-apps\endcsname{24}
\expandafter\def\csname pct@rev.desc.d6.uiv.p2.T.box.center-d1.boot-resid.n-apps\endcsname{2400.0}
\expandafter\def\csname ppt@rev.desc.d6.uiv.p2.T.box.center-d1.boot-resid.n-apps\endcsname{$+$2400.0}
\expandafter\def\csname val@rev.desc.d6.uiv.p2.T.box.center-d1.boot-resid.n-boot-valid\endcsname{4000}
\expandafter\def\csname cnt@rev.desc.d6.uiv.p2.T.box.center-d1.boot-resid.n-boot-valid\endcsname{4{,}000}
\expandafter\def\csname pct@rev.desc.d6.uiv.p2.T.box.center-d1.boot-resid.n-boot-valid\endcsname{400000.0}
\expandafter\def\csname ppt@rev.desc.d6.uiv.p2.T.box.center-d1.boot-resid.n-boot-valid\endcsname{$+$400000.0}
\expandafter\def\csname val@rev.desc.d6.uiv.p2.T.box.center-d1.boot-resid.value\endcsname{-0.009}
\expandafter\def\csname cnt@rev.desc.d6.uiv.p2.T.box.center-d1.boot-resid.value\endcsname{0}
\expandafter\def\csname pct@rev.desc.d6.uiv.p2.T.box.center-d1.boot-resid.value\endcsname{$-$0.9}
\expandafter\def\csname ppt@rev.desc.d6.uiv.p2.T.box.center-d1.boot-resid.value\endcsname{$-$0.9}
\expandafter\def\csname val@rev.desc.d6.uiv.p2.T.box.center-d1.boot-share.ci95.0\endcsname{0.694}
\expandafter\def\csname cnt@rev.desc.d6.uiv.p2.T.box.center-d1.boot-share.ci95.0\endcsname{1}
\expandafter\def\csname pct@rev.desc.d6.uiv.p2.T.box.center-d1.boot-share.ci95.0\endcsname{69.4}
\expandafter\def\csname ppt@rev.desc.d6.uiv.p2.T.box.center-d1.boot-share.ci95.0\endcsname{$+$69.4}
\expandafter\def\csname val@rev.desc.d6.uiv.p2.T.box.center-d1.boot-share.ci95.1\endcsname{0.847}
\expandafter\def\csname cnt@rev.desc.d6.uiv.p2.T.box.center-d1.boot-share.ci95.1\endcsname{1}
\expandafter\def\csname pct@rev.desc.d6.uiv.p2.T.box.center-d1.boot-share.ci95.1\endcsname{84.7}
\expandafter\def\csname ppt@rev.desc.d6.uiv.p2.T.box.center-d1.boot-share.ci95.1\endcsname{$+$84.7}
\expandafter\def\csname val@rev.desc.d6.uiv.p2.T.box.center-d1.boot-share.n-apps\endcsname{24}
\expandafter\def\csname cnt@rev.desc.d6.uiv.p2.T.box.center-d1.boot-share.n-apps\endcsname{24}
\expandafter\def\csname pct@rev.desc.d6.uiv.p2.T.box.center-d1.boot-share.n-apps\endcsname{2400.0}
\expandafter\def\csname ppt@rev.desc.d6.uiv.p2.T.box.center-d1.boot-share.n-apps\endcsname{$+$2400.0}
\expandafter\def\csname val@rev.desc.d6.uiv.p2.T.box.center-d1.boot-share.n-boot-valid\endcsname{4000}
\expandafter\def\csname cnt@rev.desc.d6.uiv.p2.T.box.center-d1.boot-share.n-boot-valid\endcsname{4{,}000}
\expandafter\def\csname pct@rev.desc.d6.uiv.p2.T.box.center-d1.boot-share.n-boot-valid\endcsname{400000.0}
\expandafter\def\csname ppt@rev.desc.d6.uiv.p2.T.box.center-d1.boot-share.n-boot-valid\endcsname{$+$400000.0}
\expandafter\def\csname val@rev.desc.d6.uiv.p2.T.box.center-d1.boot-share.value\endcsname{0.771}
\expandafter\def\csname cnt@rev.desc.d6.uiv.p2.T.box.center-d1.boot-share.value\endcsname{1}
\expandafter\def\csname pct@rev.desc.d6.uiv.p2.T.box.center-d1.boot-share.value\endcsname{77.1}
\expandafter\def\csname ppt@rev.desc.d6.uiv.p2.T.box.center-d1.boot-share.value\endcsname{$+$77.1}
\expandafter\def\csname val@rev.desc.d6.uiv.p2.T.box.center-d1.n\endcsname{144}
\expandafter\def\csname cnt@rev.desc.d6.uiv.p2.T.box.center-d1.n\endcsname{144}
\expandafter\def\csname pct@rev.desc.d6.uiv.p2.T.box.center-d1.n\endcsname{14400.0}
\expandafter\def\csname ppt@rev.desc.d6.uiv.p2.T.box.center-d1.n\endcsname{$+$14400.0}
\expandafter\def\csname val@rev.desc.d6.uiv.p2.T.box.center-d1.resid-agree-stable\endcsname{-0.009}
\expandafter\def\csname cnt@rev.desc.d6.uiv.p2.T.box.center-d1.resid-agree-stable\endcsname{0}
\expandafter\def\csname pct@rev.desc.d6.uiv.p2.T.box.center-d1.resid-agree-stable\endcsname{$-$0.9}
\expandafter\def\csname ppt@rev.desc.d6.uiv.p2.T.box.center-d1.resid-agree-stable\endcsname{$-$0.9}
\expandafter\def\csname val@rev.desc.d6.uiv.p2.T.box.center-d1.resid-disagree\endcsname{0.000}
\expandafter\def\csname cnt@rev.desc.d6.uiv.p2.T.box.center-d1.resid-disagree\endcsname{0}
\expandafter\def\csname pct@rev.desc.d6.uiv.p2.T.box.center-d1.resid-disagree\endcsname{0.0}
\expandafter\def\csname ppt@rev.desc.d6.uiv.p2.T.box.center-d1.resid-disagree\endcsname{0.0}
\expandafter\def\csname val@rev.desc.d6.uiv.p2.T.box.center-d1.share-agree-stable\endcsname{0.771}
\expandafter\def\csname cnt@rev.desc.d6.uiv.p2.T.box.center-d1.share-agree-stable\endcsname{1}
\expandafter\def\csname pct@rev.desc.d6.uiv.p2.T.box.center-d1.share-agree-stable\endcsname{77.1}
\expandafter\def\csname ppt@rev.desc.d6.uiv.p2.T.box.center-d1.share-agree-stable\endcsname{$+$77.1}
\expandafter\def\csname val@rev.desc.d6.uiv.p2.T.box.context-d1.N-agree-stable\endcsname{113}
\expandafter\def\csname cnt@rev.desc.d6.uiv.p2.T.box.context-d1.N-agree-stable\endcsname{113}
\expandafter\def\csname pct@rev.desc.d6.uiv.p2.T.box.context-d1.N-agree-stable\endcsname{11300.0}
\expandafter\def\csname ppt@rev.desc.d6.uiv.p2.T.box.context-d1.N-agree-stable\endcsname{$+$11300.0}
\expandafter\def\csname val@rev.desc.d6.uiv.p2.T.box.context-d1.N-disagree\endcsname{29}
\expandafter\def\csname cnt@rev.desc.d6.uiv.p2.T.box.context-d1.N-disagree\endcsname{29}
\expandafter\def\csname pct@rev.desc.d6.uiv.p2.T.box.context-d1.N-disagree\endcsname{2900.0}
\expandafter\def\csname ppt@rev.desc.d6.uiv.p2.T.box.context-d1.N-disagree\endcsname{$+$2900.0}
\expandafter\def\csname val@rev.desc.d6.uiv.p2.T.box.context-d1.boot-resid.ci95.0\endcsname{-0.028}
\expandafter\def\csname cnt@rev.desc.d6.uiv.p2.T.box.context-d1.boot-resid.ci95.0\endcsname{0}
\expandafter\def\csname pct@rev.desc.d6.uiv.p2.T.box.context-d1.boot-resid.ci95.0\endcsname{$-$2.8}
\expandafter\def\csname ppt@rev.desc.d6.uiv.p2.T.box.context-d1.boot-resid.ci95.0\endcsname{$-$2.8}
\expandafter\def\csname val@rev.desc.d6.uiv.p2.T.box.context-d1.boot-resid.ci95.1\endcsname{0.000}
\expandafter\def\csname cnt@rev.desc.d6.uiv.p2.T.box.context-d1.boot-resid.ci95.1\endcsname{0}
\expandafter\def\csname pct@rev.desc.d6.uiv.p2.T.box.context-d1.boot-resid.ci95.1\endcsname{0.0}
\expandafter\def\csname ppt@rev.desc.d6.uiv.p2.T.box.context-d1.boot-resid.ci95.1\endcsname{0.0}
\expandafter\def\csname val@rev.desc.d6.uiv.p2.T.box.context-d1.boot-resid.n-apps\endcsname{24}
\expandafter\def\csname cnt@rev.desc.d6.uiv.p2.T.box.context-d1.boot-resid.n-apps\endcsname{24}
\expandafter\def\csname pct@rev.desc.d6.uiv.p2.T.box.context-d1.boot-resid.n-apps\endcsname{2400.0}
\expandafter\def\csname ppt@rev.desc.d6.uiv.p2.T.box.context-d1.boot-resid.n-apps\endcsname{$+$2400.0}
\expandafter\def\csname val@rev.desc.d6.uiv.p2.T.box.context-d1.boot-resid.n-boot-valid\endcsname{4000}
\expandafter\def\csname cnt@rev.desc.d6.uiv.p2.T.box.context-d1.boot-resid.n-boot-valid\endcsname{4{,}000}
\expandafter\def\csname pct@rev.desc.d6.uiv.p2.T.box.context-d1.boot-resid.n-boot-valid\endcsname{400000.0}
\expandafter\def\csname ppt@rev.desc.d6.uiv.p2.T.box.context-d1.boot-resid.n-boot-valid\endcsname{$+$400000.0}
\expandafter\def\csname val@rev.desc.d6.uiv.p2.T.box.context-d1.boot-resid.value\endcsname{-0.009}
\expandafter\def\csname cnt@rev.desc.d6.uiv.p2.T.box.context-d1.boot-resid.value\endcsname{0}
\expandafter\def\csname pct@rev.desc.d6.uiv.p2.T.box.context-d1.boot-resid.value\endcsname{$-$0.9}
\expandafter\def\csname ppt@rev.desc.d6.uiv.p2.T.box.context-d1.boot-resid.value\endcsname{$-$0.9}
\expandafter\def\csname val@rev.desc.d6.uiv.p2.T.box.context-d1.boot-share.ci95.0\endcsname{0.722}
\expandafter\def\csname cnt@rev.desc.d6.uiv.p2.T.box.context-d1.boot-share.ci95.0\endcsname{1}
\expandafter\def\csname pct@rev.desc.d6.uiv.p2.T.box.context-d1.boot-share.ci95.0\endcsname{72.2}
\expandafter\def\csname ppt@rev.desc.d6.uiv.p2.T.box.context-d1.boot-share.ci95.0\endcsname{$+$72.2}
\expandafter\def\csname val@rev.desc.d6.uiv.p2.T.box.context-d1.boot-share.ci95.1\endcsname{0.847}
\expandafter\def\csname cnt@rev.desc.d6.uiv.p2.T.box.context-d1.boot-share.ci95.1\endcsname{1}
\expandafter\def\csname pct@rev.desc.d6.uiv.p2.T.box.context-d1.boot-share.ci95.1\endcsname{84.7}
\expandafter\def\csname ppt@rev.desc.d6.uiv.p2.T.box.context-d1.boot-share.ci95.1\endcsname{$+$84.7}
\expandafter\def\csname val@rev.desc.d6.uiv.p2.T.box.context-d1.boot-share.n-apps\endcsname{24}
\expandafter\def\csname cnt@rev.desc.d6.uiv.p2.T.box.context-d1.boot-share.n-apps\endcsname{24}
\expandafter\def\csname pct@rev.desc.d6.uiv.p2.T.box.context-d1.boot-share.n-apps\endcsname{2400.0}
\expandafter\def\csname ppt@rev.desc.d6.uiv.p2.T.box.context-d1.boot-share.n-apps\endcsname{$+$2400.0}
\expandafter\def\csname val@rev.desc.d6.uiv.p2.T.box.context-d1.boot-share.n-boot-valid\endcsname{4000}
\expandafter\def\csname cnt@rev.desc.d6.uiv.p2.T.box.context-d1.boot-share.n-boot-valid\endcsname{4{,}000}
\expandafter\def\csname pct@rev.desc.d6.uiv.p2.T.box.context-d1.boot-share.n-boot-valid\endcsname{400000.0}
\expandafter\def\csname ppt@rev.desc.d6.uiv.p2.T.box.context-d1.boot-share.n-boot-valid\endcsname{$+$400000.0}
\expandafter\def\csname val@rev.desc.d6.uiv.p2.T.box.context-d1.boot-share.value\endcsname{0.785}
\expandafter\def\csname cnt@rev.desc.d6.uiv.p2.T.box.context-d1.boot-share.value\endcsname{1}
\expandafter\def\csname pct@rev.desc.d6.uiv.p2.T.box.context-d1.boot-share.value\endcsname{78.5}
\expandafter\def\csname ppt@rev.desc.d6.uiv.p2.T.box.context-d1.boot-share.value\endcsname{$+$78.5}
\expandafter\def\csname val@rev.desc.d6.uiv.p2.T.box.context-d1.n\endcsname{144}
\expandafter\def\csname cnt@rev.desc.d6.uiv.p2.T.box.context-d1.n\endcsname{144}
\expandafter\def\csname pct@rev.desc.d6.uiv.p2.T.box.context-d1.n\endcsname{14400.0}
\expandafter\def\csname ppt@rev.desc.d6.uiv.p2.T.box.context-d1.n\endcsname{$+$14400.0}
\expandafter\def\csname val@rev.desc.d6.uiv.p2.T.box.context-d1.resid-agree-stable\endcsname{-0.009}
\expandafter\def\csname cnt@rev.desc.d6.uiv.p2.T.box.context-d1.resid-agree-stable\endcsname{0}
\expandafter\def\csname pct@rev.desc.d6.uiv.p2.T.box.context-d1.resid-agree-stable\endcsname{$-$0.9}
\expandafter\def\csname ppt@rev.desc.d6.uiv.p2.T.box.context-d1.resid-agree-stable\endcsname{$-$0.9}
\expandafter\def\csname val@rev.desc.d6.uiv.p2.T.box.context-d1.resid-disagree\endcsname{0.172}
\expandafter\def\csname cnt@rev.desc.d6.uiv.p2.T.box.context-d1.resid-disagree\endcsname{0}
\expandafter\def\csname pct@rev.desc.d6.uiv.p2.T.box.context-d1.resid-disagree\endcsname{17.2}
\expandafter\def\csname ppt@rev.desc.d6.uiv.p2.T.box.context-d1.resid-disagree\endcsname{$+$17.2}
\expandafter\def\csname val@rev.desc.d6.uiv.p2.T.box.context-d1.share-agree-stable\endcsname{0.785}
\expandafter\def\csname cnt@rev.desc.d6.uiv.p2.T.box.context-d1.share-agree-stable\endcsname{1}
\expandafter\def\csname pct@rev.desc.d6.uiv.p2.T.box.context-d1.share-agree-stable\endcsname{78.5}
\expandafter\def\csname ppt@rev.desc.d6.uiv.p2.T.box.context-d1.share-agree-stable\endcsname{$+$78.5}
\expandafter\def\csname val@rev.desc.d6.uiv.p2.T.box.dual-d2.N-agree-stable\endcsname{108}
\expandafter\def\csname cnt@rev.desc.d6.uiv.p2.T.box.dual-d2.N-agree-stable\endcsname{108}
\expandafter\def\csname pct@rev.desc.d6.uiv.p2.T.box.dual-d2.N-agree-stable\endcsname{10800.0}
\expandafter\def\csname ppt@rev.desc.d6.uiv.p2.T.box.dual-d2.N-agree-stable\endcsname{$+$10800.0}
\expandafter\def\csname val@rev.desc.d6.uiv.p2.T.box.dual-d2.N-disagree\endcsname{34}
\expandafter\def\csname cnt@rev.desc.d6.uiv.p2.T.box.dual-d2.N-disagree\endcsname{34}
\expandafter\def\csname pct@rev.desc.d6.uiv.p2.T.box.dual-d2.N-disagree\endcsname{3400.0}
\expandafter\def\csname ppt@rev.desc.d6.uiv.p2.T.box.dual-d2.N-disagree\endcsname{$+$3400.0}
\expandafter\def\csname val@rev.desc.d6.uiv.p2.T.box.dual-d2.boot-resid.ci95.0\endcsname{-0.106}
\expandafter\def\csname cnt@rev.desc.d6.uiv.p2.T.box.dual-d2.boot-resid.ci95.0\endcsname{0}
\expandafter\def\csname pct@rev.desc.d6.uiv.p2.T.box.dual-d2.boot-resid.ci95.0\endcsname{$-$10.6}
\expandafter\def\csname ppt@rev.desc.d6.uiv.p2.T.box.dual-d2.boot-resid.ci95.0\endcsname{$-$10.6}
\expandafter\def\csname val@rev.desc.d6.uiv.p2.T.box.dual-d2.boot-resid.ci95.1\endcsname{-0.017}
\expandafter\def\csname cnt@rev.desc.d6.uiv.p2.T.box.dual-d2.boot-resid.ci95.1\endcsname{0}
\expandafter\def\csname pct@rev.desc.d6.uiv.p2.T.box.dual-d2.boot-resid.ci95.1\endcsname{$-$1.7}
\expandafter\def\csname ppt@rev.desc.d6.uiv.p2.T.box.dual-d2.boot-resid.ci95.1\endcsname{$-$1.7}
\expandafter\def\csname val@rev.desc.d6.uiv.p2.T.box.dual-d2.boot-resid.n-apps\endcsname{24}
\expandafter\def\csname cnt@rev.desc.d6.uiv.p2.T.box.dual-d2.boot-resid.n-apps\endcsname{24}
\expandafter\def\csname pct@rev.desc.d6.uiv.p2.T.box.dual-d2.boot-resid.n-apps\endcsname{2400.0}
\expandafter\def\csname ppt@rev.desc.d6.uiv.p2.T.box.dual-d2.boot-resid.n-apps\endcsname{$+$2400.0}
\expandafter\def\csname val@rev.desc.d6.uiv.p2.T.box.dual-d2.boot-resid.n-boot-valid\endcsname{4000}
\expandafter\def\csname cnt@rev.desc.d6.uiv.p2.T.box.dual-d2.boot-resid.n-boot-valid\endcsname{4{,}000}
\expandafter\def\csname pct@rev.desc.d6.uiv.p2.T.box.dual-d2.boot-resid.n-boot-valid\endcsname{400000.0}
\expandafter\def\csname ppt@rev.desc.d6.uiv.p2.T.box.dual-d2.boot-resid.n-boot-valid\endcsname{$+$400000.0}
\expandafter\def\csname val@rev.desc.d6.uiv.p2.T.box.dual-d2.boot-resid.value\endcsname{-0.056}
\expandafter\def\csname cnt@rev.desc.d6.uiv.p2.T.box.dual-d2.boot-resid.value\endcsname{0}
\expandafter\def\csname pct@rev.desc.d6.uiv.p2.T.box.dual-d2.boot-resid.value\endcsname{$-$5.6}
\expandafter\def\csname ppt@rev.desc.d6.uiv.p2.T.box.dual-d2.boot-resid.value\endcsname{$-$5.6}
\expandafter\def\csname val@rev.desc.d6.uiv.p2.T.box.dual-d2.boot-share.ci95.0\endcsname{0.674}
\expandafter\def\csname cnt@rev.desc.d6.uiv.p2.T.box.dual-d2.boot-share.ci95.0\endcsname{1}
\expandafter\def\csname pct@rev.desc.d6.uiv.p2.T.box.dual-d2.boot-share.ci95.0\endcsname{67.4}
\expandafter\def\csname ppt@rev.desc.d6.uiv.p2.T.box.dual-d2.boot-share.ci95.0\endcsname{$+$67.4}
\expandafter\def\csname val@rev.desc.d6.uiv.p2.T.box.dual-d2.boot-share.ci95.1\endcsname{0.826}
\expandafter\def\csname cnt@rev.desc.d6.uiv.p2.T.box.dual-d2.boot-share.ci95.1\endcsname{1}
\expandafter\def\csname pct@rev.desc.d6.uiv.p2.T.box.dual-d2.boot-share.ci95.1\endcsname{82.6}
\expandafter\def\csname ppt@rev.desc.d6.uiv.p2.T.box.dual-d2.boot-share.ci95.1\endcsname{$+$82.6}
\expandafter\def\csname val@rev.desc.d6.uiv.p2.T.box.dual-d2.boot-share.n-apps\endcsname{24}
\expandafter\def\csname cnt@rev.desc.d6.uiv.p2.T.box.dual-d2.boot-share.n-apps\endcsname{24}
\expandafter\def\csname pct@rev.desc.d6.uiv.p2.T.box.dual-d2.boot-share.n-apps\endcsname{2400.0}
\expandafter\def\csname ppt@rev.desc.d6.uiv.p2.T.box.dual-d2.boot-share.n-apps\endcsname{$+$2400.0}
\expandafter\def\csname val@rev.desc.d6.uiv.p2.T.box.dual-d2.boot-share.n-boot-valid\endcsname{4000}
\expandafter\def\csname cnt@rev.desc.d6.uiv.p2.T.box.dual-d2.boot-share.n-boot-valid\endcsname{4{,}000}
\expandafter\def\csname pct@rev.desc.d6.uiv.p2.T.box.dual-d2.boot-share.n-boot-valid\endcsname{400000.0}
\expandafter\def\csname ppt@rev.desc.d6.uiv.p2.T.box.dual-d2.boot-share.n-boot-valid\endcsname{$+$400000.0}
\expandafter\def\csname val@rev.desc.d6.uiv.p2.T.box.dual-d2.boot-share.value\endcsname{0.750}
\expandafter\def\csname cnt@rev.desc.d6.uiv.p2.T.box.dual-d2.boot-share.value\endcsname{1}
\expandafter\def\csname pct@rev.desc.d6.uiv.p2.T.box.dual-d2.boot-share.value\endcsname{75.0}
\expandafter\def\csname ppt@rev.desc.d6.uiv.p2.T.box.dual-d2.boot-share.value\endcsname{$+$75.0}
\expandafter\def\csname val@rev.desc.d6.uiv.p2.T.box.dual-d2.n\endcsname{144}
\expandafter\def\csname cnt@rev.desc.d6.uiv.p2.T.box.dual-d2.n\endcsname{144}
\expandafter\def\csname pct@rev.desc.d6.uiv.p2.T.box.dual-d2.n\endcsname{14400.0}
\expandafter\def\csname ppt@rev.desc.d6.uiv.p2.T.box.dual-d2.n\endcsname{$+$14400.0}
\expandafter\def\csname val@rev.desc.d6.uiv.p2.T.box.dual-d2.resid-agree-stable\endcsname{-0.056}
\expandafter\def\csname cnt@rev.desc.d6.uiv.p2.T.box.dual-d2.resid-agree-stable\endcsname{0}
\expandafter\def\csname pct@rev.desc.d6.uiv.p2.T.box.dual-d2.resid-agree-stable\endcsname{$-$5.6}
\expandafter\def\csname ppt@rev.desc.d6.uiv.p2.T.box.dual-d2.resid-agree-stable\endcsname{$-$5.6}
\expandafter\def\csname val@rev.desc.d6.uiv.p2.T.box.dual-d2.resid-disagree\endcsname{0.147}
\expandafter\def\csname cnt@rev.desc.d6.uiv.p2.T.box.dual-d2.resid-disagree\endcsname{0}
\expandafter\def\csname pct@rev.desc.d6.uiv.p2.T.box.dual-d2.resid-disagree\endcsname{14.7}
\expandafter\def\csname ppt@rev.desc.d6.uiv.p2.T.box.dual-d2.resid-disagree\endcsname{$+$14.7}
\expandafter\def\csname val@rev.desc.d6.uiv.p2.T.box.dual-d2.share-agree-stable\endcsname{0.750}
\expandafter\def\csname cnt@rev.desc.d6.uiv.p2.T.box.dual-d2.share-agree-stable\endcsname{1}
\expandafter\def\csname pct@rev.desc.d6.uiv.p2.T.box.dual-d2.share-agree-stable\endcsname{75.0}
\expandafter\def\csname ppt@rev.desc.d6.uiv.p2.T.box.dual-d2.share-agree-stable\endcsname{$+$75.0}
\expandafter\def\csname val@rev.desc.d6.uiv.p2.T.box.grid2-d3.N-agree-stable\endcsname{109}
\expandafter\def\csname cnt@rev.desc.d6.uiv.p2.T.box.grid2-d3.N-agree-stable\endcsname{109}
\expandafter\def\csname pct@rev.desc.d6.uiv.p2.T.box.grid2-d3.N-agree-stable\endcsname{10900.0}
\expandafter\def\csname ppt@rev.desc.d6.uiv.p2.T.box.grid2-d3.N-agree-stable\endcsname{$+$10900.0}
\expandafter\def\csname val@rev.desc.d6.uiv.p2.T.box.grid2-d3.N-disagree\endcsname{31}
\expandafter\def\csname cnt@rev.desc.d6.uiv.p2.T.box.grid2-d3.N-disagree\endcsname{31}
\expandafter\def\csname pct@rev.desc.d6.uiv.p2.T.box.grid2-d3.N-disagree\endcsname{3100.0}
\expandafter\def\csname ppt@rev.desc.d6.uiv.p2.T.box.grid2-d3.N-disagree\endcsname{$+$3100.0}
\expandafter\def\csname val@rev.desc.d6.uiv.p2.T.box.grid2-d3.boot-resid.ci95.0\endcsname{-0.048}
\expandafter\def\csname cnt@rev.desc.d6.uiv.p2.T.box.grid2-d3.boot-resid.ci95.0\endcsname{0}
\expandafter\def\csname pct@rev.desc.d6.uiv.p2.T.box.grid2-d3.boot-resid.ci95.0\endcsname{$-$4.8}
\expandafter\def\csname ppt@rev.desc.d6.uiv.p2.T.box.grid2-d3.boot-resid.ci95.0\endcsname{$-$4.8}
\expandafter\def\csname val@rev.desc.d6.uiv.p2.T.box.grid2-d3.boot-resid.ci95.1\endcsname{0.000}
\expandafter\def\csname cnt@rev.desc.d6.uiv.p2.T.box.grid2-d3.boot-resid.ci95.1\endcsname{0}
\expandafter\def\csname pct@rev.desc.d6.uiv.p2.T.box.grid2-d3.boot-resid.ci95.1\endcsname{0.0}
\expandafter\def\csname ppt@rev.desc.d6.uiv.p2.T.box.grid2-d3.boot-resid.ci95.1\endcsname{0.0}
\expandafter\def\csname val@rev.desc.d6.uiv.p2.T.box.grid2-d3.boot-resid.n-apps\endcsname{24}
\expandafter\def\csname cnt@rev.desc.d6.uiv.p2.T.box.grid2-d3.boot-resid.n-apps\endcsname{24}
\expandafter\def\csname pct@rev.desc.d6.uiv.p2.T.box.grid2-d3.boot-resid.n-apps\endcsname{2400.0}
\expandafter\def\csname ppt@rev.desc.d6.uiv.p2.T.box.grid2-d3.boot-resid.n-apps\endcsname{$+$2400.0}
\expandafter\def\csname val@rev.desc.d6.uiv.p2.T.box.grid2-d3.boot-resid.n-boot-valid\endcsname{4000}
\expandafter\def\csname cnt@rev.desc.d6.uiv.p2.T.box.grid2-d3.boot-resid.n-boot-valid\endcsname{4{,}000}
\expandafter\def\csname pct@rev.desc.d6.uiv.p2.T.box.grid2-d3.boot-resid.n-boot-valid\endcsname{400000.0}
\expandafter\def\csname ppt@rev.desc.d6.uiv.p2.T.box.grid2-d3.boot-resid.n-boot-valid\endcsname{$+$400000.0}
\expandafter\def\csname val@rev.desc.d6.uiv.p2.T.box.grid2-d3.boot-resid.value\endcsname{-0.018}
\expandafter\def\csname cnt@rev.desc.d6.uiv.p2.T.box.grid2-d3.boot-resid.value\endcsname{0}
\expandafter\def\csname pct@rev.desc.d6.uiv.p2.T.box.grid2-d3.boot-resid.value\endcsname{$-$1.8}
\expandafter\def\csname ppt@rev.desc.d6.uiv.p2.T.box.grid2-d3.boot-resid.value\endcsname{$-$1.8}
\expandafter\def\csname val@rev.desc.d6.uiv.p2.T.box.grid2-d3.boot-share.ci95.0\endcsname{0.688}
\expandafter\def\csname cnt@rev.desc.d6.uiv.p2.T.box.grid2-d3.boot-share.ci95.0\endcsname{1}
\expandafter\def\csname pct@rev.desc.d6.uiv.p2.T.box.grid2-d3.boot-share.ci95.0\endcsname{68.8}
\expandafter\def\csname ppt@rev.desc.d6.uiv.p2.T.box.grid2-d3.boot-share.ci95.0\endcsname{$+$68.8}
\expandafter\def\csname val@rev.desc.d6.uiv.p2.T.box.grid2-d3.boot-share.ci95.1\endcsname{0.833}
\expandafter\def\csname cnt@rev.desc.d6.uiv.p2.T.box.grid2-d3.boot-share.ci95.1\endcsname{1}
\expandafter\def\csname pct@rev.desc.d6.uiv.p2.T.box.grid2-d3.boot-share.ci95.1\endcsname{83.3}
\expandafter\def\csname ppt@rev.desc.d6.uiv.p2.T.box.grid2-d3.boot-share.ci95.1\endcsname{$+$83.3}
\expandafter\def\csname val@rev.desc.d6.uiv.p2.T.box.grid2-d3.boot-share.n-apps\endcsname{24}
\expandafter\def\csname cnt@rev.desc.d6.uiv.p2.T.box.grid2-d3.boot-share.n-apps\endcsname{24}
\expandafter\def\csname pct@rev.desc.d6.uiv.p2.T.box.grid2-d3.boot-share.n-apps\endcsname{2400.0}
\expandafter\def\csname ppt@rev.desc.d6.uiv.p2.T.box.grid2-d3.boot-share.n-apps\endcsname{$+$2400.0}
\expandafter\def\csname val@rev.desc.d6.uiv.p2.T.box.grid2-d3.boot-share.n-boot-valid\endcsname{4000}
\expandafter\def\csname cnt@rev.desc.d6.uiv.p2.T.box.grid2-d3.boot-share.n-boot-valid\endcsname{4{,}000}
\expandafter\def\csname pct@rev.desc.d6.uiv.p2.T.box.grid2-d3.boot-share.n-boot-valid\endcsname{400000.0}
\expandafter\def\csname ppt@rev.desc.d6.uiv.p2.T.box.grid2-d3.boot-share.n-boot-valid\endcsname{$+$400000.0}
\expandafter\def\csname val@rev.desc.d6.uiv.p2.T.box.grid2-d3.boot-share.value\endcsname{0.757}
\expandafter\def\csname cnt@rev.desc.d6.uiv.p2.T.box.grid2-d3.boot-share.value\endcsname{1}
\expandafter\def\csname pct@rev.desc.d6.uiv.p2.T.box.grid2-d3.boot-share.value\endcsname{75.7}
\expandafter\def\csname ppt@rev.desc.d6.uiv.p2.T.box.grid2-d3.boot-share.value\endcsname{$+$75.7}
\expandafter\def\csname val@rev.desc.d6.uiv.p2.T.box.grid2-d3.n\endcsname{144}
\expandafter\def\csname cnt@rev.desc.d6.uiv.p2.T.box.grid2-d3.n\endcsname{144}
\expandafter\def\csname pct@rev.desc.d6.uiv.p2.T.box.grid2-d3.n\endcsname{14400.0}
\expandafter\def\csname ppt@rev.desc.d6.uiv.p2.T.box.grid2-d3.n\endcsname{$+$14400.0}
\expandafter\def\csname val@rev.desc.d6.uiv.p2.T.box.grid2-d3.resid-agree-stable\endcsname{-0.018}
\expandafter\def\csname cnt@rev.desc.d6.uiv.p2.T.box.grid2-d3.resid-agree-stable\endcsname{0}
\expandafter\def\csname pct@rev.desc.d6.uiv.p2.T.box.grid2-d3.resid-agree-stable\endcsname{$-$1.8}
\expandafter\def\csname ppt@rev.desc.d6.uiv.p2.T.box.grid2-d3.resid-agree-stable\endcsname{$-$1.8}
\expandafter\def\csname val@rev.desc.d6.uiv.p2.T.box.grid2-d3.resid-disagree\endcsname{0.194}
\expandafter\def\csname cnt@rev.desc.d6.uiv.p2.T.box.grid2-d3.resid-disagree\endcsname{0}
\expandafter\def\csname pct@rev.desc.d6.uiv.p2.T.box.grid2-d3.resid-disagree\endcsname{19.4}
\expandafter\def\csname ppt@rev.desc.d6.uiv.p2.T.box.grid2-d3.resid-disagree\endcsname{$+$19.4}
\expandafter\def\csname val@rev.desc.d6.uiv.p2.T.box.grid2-d3.share-agree-stable\endcsname{0.757}
\expandafter\def\csname cnt@rev.desc.d6.uiv.p2.T.box.grid2-d3.share-agree-stable\endcsname{1}
\expandafter\def\csname pct@rev.desc.d6.uiv.p2.T.box.grid2-d3.share-agree-stable\endcsname{75.7}
\expandafter\def\csname ppt@rev.desc.d6.uiv.p2.T.box.grid2-d3.share-agree-stable\endcsname{$+$75.7}
\expandafter\def\csname val@rev.desc.d6.uiv.p2.T.box.zoomclick-d3.N-agree-stable\endcsname{111}
\expandafter\def\csname cnt@rev.desc.d6.uiv.p2.T.box.zoomclick-d3.N-agree-stable\endcsname{111}
\expandafter\def\csname pct@rev.desc.d6.uiv.p2.T.box.zoomclick-d3.N-agree-stable\endcsname{11100.0}
\expandafter\def\csname ppt@rev.desc.d6.uiv.p2.T.box.zoomclick-d3.N-agree-stable\endcsname{$+$11100.0}
\expandafter\def\csname val@rev.desc.d6.uiv.p2.T.box.zoomclick-d3.N-disagree\endcsname{29}
\expandafter\def\csname cnt@rev.desc.d6.uiv.p2.T.box.zoomclick-d3.N-disagree\endcsname{29}
\expandafter\def\csname pct@rev.desc.d6.uiv.p2.T.box.zoomclick-d3.N-disagree\endcsname{2900.0}
\expandafter\def\csname ppt@rev.desc.d6.uiv.p2.T.box.zoomclick-d3.N-disagree\endcsname{$+$2900.0}
\expandafter\def\csname val@rev.desc.d6.uiv.p2.T.box.zoomclick-d3.boot-resid.ci95.0\endcsname{-0.028}
\expandafter\def\csname cnt@rev.desc.d6.uiv.p2.T.box.zoomclick-d3.boot-resid.ci95.0\endcsname{0}
\expandafter\def\csname pct@rev.desc.d6.uiv.p2.T.box.zoomclick-d3.boot-resid.ci95.0\endcsname{$-$2.8}
\expandafter\def\csname ppt@rev.desc.d6.uiv.p2.T.box.zoomclick-d3.boot-resid.ci95.0\endcsname{$-$2.8}
\expandafter\def\csname val@rev.desc.d6.uiv.p2.T.box.zoomclick-d3.boot-resid.ci95.1\endcsname{0.000}
\expandafter\def\csname cnt@rev.desc.d6.uiv.p2.T.box.zoomclick-d3.boot-resid.ci95.1\endcsname{0}
\expandafter\def\csname pct@rev.desc.d6.uiv.p2.T.box.zoomclick-d3.boot-resid.ci95.1\endcsname{0.0}
\expandafter\def\csname ppt@rev.desc.d6.uiv.p2.T.box.zoomclick-d3.boot-resid.ci95.1\endcsname{0.0}
\expandafter\def\csname val@rev.desc.d6.uiv.p2.T.box.zoomclick-d3.boot-resid.n-apps\endcsname{24}
\expandafter\def\csname cnt@rev.desc.d6.uiv.p2.T.box.zoomclick-d3.boot-resid.n-apps\endcsname{24}
\expandafter\def\csname pct@rev.desc.d6.uiv.p2.T.box.zoomclick-d3.boot-resid.n-apps\endcsname{2400.0}
\expandafter\def\csname ppt@rev.desc.d6.uiv.p2.T.box.zoomclick-d3.boot-resid.n-apps\endcsname{$+$2400.0}
\expandafter\def\csname val@rev.desc.d6.uiv.p2.T.box.zoomclick-d3.boot-resid.n-boot-valid\endcsname{4000}
\expandafter\def\csname cnt@rev.desc.d6.uiv.p2.T.box.zoomclick-d3.boot-resid.n-boot-valid\endcsname{4{,}000}
\expandafter\def\csname pct@rev.desc.d6.uiv.p2.T.box.zoomclick-d3.boot-resid.n-boot-valid\endcsname{400000.0}
\expandafter\def\csname ppt@rev.desc.d6.uiv.p2.T.box.zoomclick-d3.boot-resid.n-boot-valid\endcsname{$+$400000.0}
\expandafter\def\csname val@rev.desc.d6.uiv.p2.T.box.zoomclick-d3.boot-resid.value\endcsname{-0.009}
\expandafter\def\csname cnt@rev.desc.d6.uiv.p2.T.box.zoomclick-d3.boot-resid.value\endcsname{0}
\expandafter\def\csname pct@rev.desc.d6.uiv.p2.T.box.zoomclick-d3.boot-resid.value\endcsname{$-$0.9}
\expandafter\def\csname ppt@rev.desc.d6.uiv.p2.T.box.zoomclick-d3.boot-resid.value\endcsname{$-$0.9}
\expandafter\def\csname val@rev.desc.d6.uiv.p2.T.box.zoomclick-d3.boot-share.ci95.0\endcsname{0.694}
\expandafter\def\csname cnt@rev.desc.d6.uiv.p2.T.box.zoomclick-d3.boot-share.ci95.0\endcsname{1}
\expandafter\def\csname pct@rev.desc.d6.uiv.p2.T.box.zoomclick-d3.boot-share.ci95.0\endcsname{69.4}
\expandafter\def\csname ppt@rev.desc.d6.uiv.p2.T.box.zoomclick-d3.boot-share.ci95.0\endcsname{$+$69.4}
\expandafter\def\csname val@rev.desc.d6.uiv.p2.T.box.zoomclick-d3.boot-share.ci95.1\endcsname{0.847}
\expandafter\def\csname cnt@rev.desc.d6.uiv.p2.T.box.zoomclick-d3.boot-share.ci95.1\endcsname{1}
\expandafter\def\csname pct@rev.desc.d6.uiv.p2.T.box.zoomclick-d3.boot-share.ci95.1\endcsname{84.7}
\expandafter\def\csname ppt@rev.desc.d6.uiv.p2.T.box.zoomclick-d3.boot-share.ci95.1\endcsname{$+$84.7}
\expandafter\def\csname val@rev.desc.d6.uiv.p2.T.box.zoomclick-d3.boot-share.n-apps\endcsname{24}
\expandafter\def\csname cnt@rev.desc.d6.uiv.p2.T.box.zoomclick-d3.boot-share.n-apps\endcsname{24}
\expandafter\def\csname pct@rev.desc.d6.uiv.p2.T.box.zoomclick-d3.boot-share.n-apps\endcsname{2400.0}
\expandafter\def\csname ppt@rev.desc.d6.uiv.p2.T.box.zoomclick-d3.boot-share.n-apps\endcsname{$+$2400.0}
\expandafter\def\csname val@rev.desc.d6.uiv.p2.T.box.zoomclick-d3.boot-share.n-boot-valid\endcsname{4000}
\expandafter\def\csname cnt@rev.desc.d6.uiv.p2.T.box.zoomclick-d3.boot-share.n-boot-valid\endcsname{4{,}000}
\expandafter\def\csname pct@rev.desc.d6.uiv.p2.T.box.zoomclick-d3.boot-share.n-boot-valid\endcsname{400000.0}
\expandafter\def\csname ppt@rev.desc.d6.uiv.p2.T.box.zoomclick-d3.boot-share.n-boot-valid\endcsname{$+$400000.0}
\expandafter\def\csname val@rev.desc.d6.uiv.p2.T.box.zoomclick-d3.boot-share.value\endcsname{0.771}
\expandafter\def\csname cnt@rev.desc.d6.uiv.p2.T.box.zoomclick-d3.boot-share.value\endcsname{1}
\expandafter\def\csname pct@rev.desc.d6.uiv.p2.T.box.zoomclick-d3.boot-share.value\endcsname{77.1}
\expandafter\def\csname ppt@rev.desc.d6.uiv.p2.T.box.zoomclick-d3.boot-share.value\endcsname{$+$77.1}
\expandafter\def\csname val@rev.desc.d6.uiv.p2.T.box.zoomclick-d3.n\endcsname{144}
\expandafter\def\csname cnt@rev.desc.d6.uiv.p2.T.box.zoomclick-d3.n\endcsname{144}
\expandafter\def\csname pct@rev.desc.d6.uiv.p2.T.box.zoomclick-d3.n\endcsname{14400.0}
\expandafter\def\csname ppt@rev.desc.d6.uiv.p2.T.box.zoomclick-d3.n\endcsname{$+$14400.0}
\expandafter\def\csname val@rev.desc.d6.uiv.p2.T.box.zoomclick-d3.resid-agree-stable\endcsname{-0.009}
\expandafter\def\csname cnt@rev.desc.d6.uiv.p2.T.box.zoomclick-d3.resid-agree-stable\endcsname{0}
\expandafter\def\csname pct@rev.desc.d6.uiv.p2.T.box.zoomclick-d3.resid-agree-stable\endcsname{$-$0.9}
\expandafter\def\csname ppt@rev.desc.d6.uiv.p2.T.box.zoomclick-d3.resid-agree-stable\endcsname{$-$0.9}
\expandafter\def\csname val@rev.desc.d6.uiv.p2.T.box.zoomclick-d3.resid-disagree\endcsname{0.103}
\expandafter\def\csname cnt@rev.desc.d6.uiv.p2.T.box.zoomclick-d3.resid-disagree\endcsname{0}
\expandafter\def\csname pct@rev.desc.d6.uiv.p2.T.box.zoomclick-d3.resid-disagree\endcsname{10.3}
\expandafter\def\csname ppt@rev.desc.d6.uiv.p2.T.box.zoomclick-d3.resid-disagree\endcsname{$+$10.3}
\expandafter\def\csname val@rev.desc.d6.uiv.p2.T.box.zoomclick-d3.share-agree-stable\endcsname{0.771}
\expandafter\def\csname cnt@rev.desc.d6.uiv.p2.T.box.zoomclick-d3.share-agree-stable\endcsname{1}
\expandafter\def\csname pct@rev.desc.d6.uiv.p2.T.box.zoomclick-d3.share-agree-stable\endcsname{77.1}
\expandafter\def\csname ppt@rev.desc.d6.uiv.p2.T.box.zoomclick-d3.share-agree-stable\endcsname{$+$77.1}
\expandafter\def\csname val@rev.desc.d6.uiv.p2.T.tdiag.catalog-pooled.N-agree-stable\endcsname{516}
\expandafter\def\csname cnt@rev.desc.d6.uiv.p2.T.tdiag.catalog-pooled.N-agree-stable\endcsname{516}
\expandafter\def\csname pct@rev.desc.d6.uiv.p2.T.tdiag.catalog-pooled.N-agree-stable\endcsname{51600.0}
\expandafter\def\csname ppt@rev.desc.d6.uiv.p2.T.tdiag.catalog-pooled.N-agree-stable\endcsname{$+$51600.0}
\expandafter\def\csname val@rev.desc.d6.uiv.p2.T.tdiag.catalog-pooled.N-disagree\endcsname{43}
\expandafter\def\csname cnt@rev.desc.d6.uiv.p2.T.tdiag.catalog-pooled.N-disagree\endcsname{43}
\expandafter\def\csname pct@rev.desc.d6.uiv.p2.T.tdiag.catalog-pooled.N-disagree\endcsname{4300.0}
\expandafter\def\csname ppt@rev.desc.d6.uiv.p2.T.tdiag.catalog-pooled.N-disagree\endcsname{$+$4300.0}
\expandafter\def\csname val@rev.desc.d6.uiv.p2.T.tdiag.catalog-pooled.boot-resid.ci95.0\endcsname{-0.040}
\expandafter\def\csname cnt@rev.desc.d6.uiv.p2.T.tdiag.catalog-pooled.boot-resid.ci95.0\endcsname{0}
\expandafter\def\csname pct@rev.desc.d6.uiv.p2.T.tdiag.catalog-pooled.boot-resid.ci95.0\endcsname{$-$4.0}
\expandafter\def\csname ppt@rev.desc.d6.uiv.p2.T.tdiag.catalog-pooled.boot-resid.ci95.0\endcsname{$-$4.0}
\expandafter\def\csname val@rev.desc.d6.uiv.p2.T.tdiag.catalog-pooled.boot-resid.ci95.1\endcsname{0.008}
\expandafter\def\csname cnt@rev.desc.d6.uiv.p2.T.tdiag.catalog-pooled.boot-resid.ci95.1\endcsname{0}
\expandafter\def\csname pct@rev.desc.d6.uiv.p2.T.tdiag.catalog-pooled.boot-resid.ci95.1\endcsname{0.8}
\expandafter\def\csname ppt@rev.desc.d6.uiv.p2.T.tdiag.catalog-pooled.boot-resid.ci95.1\endcsname{$+$0.8}
\expandafter\def\csname val@rev.desc.d6.uiv.p2.T.tdiag.catalog-pooled.boot-resid.n-apps\endcsname{24}
\expandafter\def\csname cnt@rev.desc.d6.uiv.p2.T.tdiag.catalog-pooled.boot-resid.n-apps\endcsname{24}
\expandafter\def\csname pct@rev.desc.d6.uiv.p2.T.tdiag.catalog-pooled.boot-resid.n-apps\endcsname{2400.0}
\expandafter\def\csname ppt@rev.desc.d6.uiv.p2.T.tdiag.catalog-pooled.boot-resid.n-apps\endcsname{$+$2400.0}
\expandafter\def\csname val@rev.desc.d6.uiv.p2.T.tdiag.catalog-pooled.boot-resid.n-boot-valid\endcsname{4000}
\expandafter\def\csname cnt@rev.desc.d6.uiv.p2.T.tdiag.catalog-pooled.boot-resid.n-boot-valid\endcsname{4{,}000}
\expandafter\def\csname pct@rev.desc.d6.uiv.p2.T.tdiag.catalog-pooled.boot-resid.n-boot-valid\endcsname{400000.0}
\expandafter\def\csname ppt@rev.desc.d6.uiv.p2.T.tdiag.catalog-pooled.boot-resid.n-boot-valid\endcsname{$+$400000.0}
\expandafter\def\csname val@rev.desc.d6.uiv.p2.T.tdiag.catalog-pooled.boot-resid.value\endcsname{-0.014}
\expandafter\def\csname cnt@rev.desc.d6.uiv.p2.T.tdiag.catalog-pooled.boot-resid.value\endcsname{0}
\expandafter\def\csname pct@rev.desc.d6.uiv.p2.T.tdiag.catalog-pooled.boot-resid.value\endcsname{$-$1.4}
\expandafter\def\csname ppt@rev.desc.d6.uiv.p2.T.tdiag.catalog-pooled.boot-resid.value\endcsname{$-$1.4}
\expandafter\def\csname val@rev.desc.d6.uiv.p2.T.tdiag.catalog-pooled.boot-share.ci95.0\endcsname{0.835}
\expandafter\def\csname cnt@rev.desc.d6.uiv.p2.T.tdiag.catalog-pooled.boot-share.ci95.0\endcsname{1}
\expandafter\def\csname pct@rev.desc.d6.uiv.p2.T.tdiag.catalog-pooled.boot-share.ci95.0\endcsname{83.5}
\expandafter\def\csname ppt@rev.desc.d6.uiv.p2.T.tdiag.catalog-pooled.boot-share.ci95.0\endcsname{$+$83.5}
\expandafter\def\csname val@rev.desc.d6.uiv.p2.T.tdiag.catalog-pooled.boot-share.ci95.1\endcsname{0.946}
\expandafter\def\csname cnt@rev.desc.d6.uiv.p2.T.tdiag.catalog-pooled.boot-share.ci95.1\endcsname{1}
\expandafter\def\csname pct@rev.desc.d6.uiv.p2.T.tdiag.catalog-pooled.boot-share.ci95.1\endcsname{94.6}
\expandafter\def\csname ppt@rev.desc.d6.uiv.p2.T.tdiag.catalog-pooled.boot-share.ci95.1\endcsname{$+$94.6}
\expandafter\def\csname val@rev.desc.d6.uiv.p2.T.tdiag.catalog-pooled.boot-share.n-apps\endcsname{24}
\expandafter\def\csname cnt@rev.desc.d6.uiv.p2.T.tdiag.catalog-pooled.boot-share.n-apps\endcsname{24}
\expandafter\def\csname pct@rev.desc.d6.uiv.p2.T.tdiag.catalog-pooled.boot-share.n-apps\endcsname{2400.0}
\expandafter\def\csname ppt@rev.desc.d6.uiv.p2.T.tdiag.catalog-pooled.boot-share.n-apps\endcsname{$+$2400.0}
\expandafter\def\csname val@rev.desc.d6.uiv.p2.T.tdiag.catalog-pooled.boot-share.n-boot-valid\endcsname{4000}
\expandafter\def\csname cnt@rev.desc.d6.uiv.p2.T.tdiag.catalog-pooled.boot-share.n-boot-valid\endcsname{4{,}000}
\expandafter\def\csname pct@rev.desc.d6.uiv.p2.T.tdiag.catalog-pooled.boot-share.n-boot-valid\endcsname{400000.0}
\expandafter\def\csname ppt@rev.desc.d6.uiv.p2.T.tdiag.catalog-pooled.boot-share.n-boot-valid\endcsname{$+$400000.0}
\expandafter\def\csname val@rev.desc.d6.uiv.p2.T.tdiag.catalog-pooled.boot-share.value\endcsname{0.896}
\expandafter\def\csname cnt@rev.desc.d6.uiv.p2.T.tdiag.catalog-pooled.boot-share.value\endcsname{1}
\expandafter\def\csname pct@rev.desc.d6.uiv.p2.T.tdiag.catalog-pooled.boot-share.value\endcsname{89.6}
\expandafter\def\csname ppt@rev.desc.d6.uiv.p2.T.tdiag.catalog-pooled.boot-share.value\endcsname{$+$89.6}
\expandafter\def\csname val@rev.desc.d6.uiv.p2.T.tdiag.catalog-pooled.gap\endcsname{0.293}
\expandafter\def\csname cnt@rev.desc.d6.uiv.p2.T.tdiag.catalog-pooled.gap\endcsname{0}
\expandafter\def\csname pct@rev.desc.d6.uiv.p2.T.tdiag.catalog-pooled.gap\endcsname{29.3}
\expandafter\def\csname ppt@rev.desc.d6.uiv.p2.T.tdiag.catalog-pooled.gap\endcsname{$+$29.3}
\expandafter\def\csname val@rev.desc.d6.uiv.p2.T.tdiag.catalog-pooled.n\endcsname{576}
\expandafter\def\csname cnt@rev.desc.d6.uiv.p2.T.tdiag.catalog-pooled.n\endcsname{576}
\expandafter\def\csname pct@rev.desc.d6.uiv.p2.T.tdiag.catalog-pooled.n\endcsname{57600.0}
\expandafter\def\csname ppt@rev.desc.d6.uiv.p2.T.tdiag.catalog-pooled.n\endcsname{$+$57600.0}
\expandafter\def\csname val@rev.desc.d6.uiv.p2.T.tdiag.catalog-pooled.resid-agree-stable\endcsname{-0.014}
\expandafter\def\csname cnt@rev.desc.d6.uiv.p2.T.tdiag.catalog-pooled.resid-agree-stable\endcsname{0}
\expandafter\def\csname pct@rev.desc.d6.uiv.p2.T.tdiag.catalog-pooled.resid-agree-stable\endcsname{$-$1.4}
\expandafter\def\csname ppt@rev.desc.d6.uiv.p2.T.tdiag.catalog-pooled.resid-agree-stable\endcsname{$-$1.4}
\expandafter\def\csname val@rev.desc.d6.uiv.p2.T.tdiag.catalog-pooled.resid-disagree\endcsname{0.279}
\expandafter\def\csname cnt@rev.desc.d6.uiv.p2.T.tdiag.catalog-pooled.resid-disagree\endcsname{0}
\expandafter\def\csname pct@rev.desc.d6.uiv.p2.T.tdiag.catalog-pooled.resid-disagree\endcsname{27.9}
\expandafter\def\csname ppt@rev.desc.d6.uiv.p2.T.tdiag.catalog-pooled.resid-disagree\endcsname{$+$27.9}
\expandafter\def\csname val@rev.desc.d6.uiv.p2.T.tdiag.catalog-pooled.share-agree-stable\endcsname{0.896}
\expandafter\def\csname cnt@rev.desc.d6.uiv.p2.T.tdiag.catalog-pooled.share-agree-stable\endcsname{1}
\expandafter\def\csname pct@rev.desc.d6.uiv.p2.T.tdiag.catalog-pooled.share-agree-stable\endcsname{89.6}
\expandafter\def\csname ppt@rev.desc.d6.uiv.p2.T.tdiag.catalog-pooled.share-agree-stable\endcsname{$+$89.6}
\expandafter\def\csname val@rev.desc.d6.uiv.p2.T.tdiag.center-d1.N-agree-stable\endcsname{130}
\expandafter\def\csname cnt@rev.desc.d6.uiv.p2.T.tdiag.center-d1.N-agree-stable\endcsname{130}
\expandafter\def\csname pct@rev.desc.d6.uiv.p2.T.tdiag.center-d1.N-agree-stable\endcsname{13000.0}
\expandafter\def\csname ppt@rev.desc.d6.uiv.p2.T.tdiag.center-d1.N-agree-stable\endcsname{$+$13000.0}
\expandafter\def\csname val@rev.desc.d6.uiv.p2.T.tdiag.center-d1.N-disagree\endcsname{9}
\expandafter\def\csname cnt@rev.desc.d6.uiv.p2.T.tdiag.center-d1.N-disagree\endcsname{9}
\expandafter\def\csname pct@rev.desc.d6.uiv.p2.T.tdiag.center-d1.N-disagree\endcsname{900.0}
\expandafter\def\csname ppt@rev.desc.d6.uiv.p2.T.tdiag.center-d1.N-disagree\endcsname{$+$900.0}
\expandafter\def\csname val@rev.desc.d6.uiv.p2.T.tdiag.center-d1.boot-resid.ci95.0\endcsname{-0.038}
\expandafter\def\csname cnt@rev.desc.d6.uiv.p2.T.tdiag.center-d1.boot-resid.ci95.0\endcsname{0}
\expandafter\def\csname pct@rev.desc.d6.uiv.p2.T.tdiag.center-d1.boot-resid.ci95.0\endcsname{$-$3.8}
\expandafter\def\csname ppt@rev.desc.d6.uiv.p2.T.tdiag.center-d1.boot-resid.ci95.0\endcsname{$-$3.8}
\expandafter\def\csname val@rev.desc.d6.uiv.p2.T.tdiag.center-d1.boot-resid.ci95.1\endcsname{0.000}
\expandafter\def\csname cnt@rev.desc.d6.uiv.p2.T.tdiag.center-d1.boot-resid.ci95.1\endcsname{0}
\expandafter\def\csname pct@rev.desc.d6.uiv.p2.T.tdiag.center-d1.boot-resid.ci95.1\endcsname{0.0}
\expandafter\def\csname ppt@rev.desc.d6.uiv.p2.T.tdiag.center-d1.boot-resid.ci95.1\endcsname{0.0}
\expandafter\def\csname val@rev.desc.d6.uiv.p2.T.tdiag.center-d1.boot-resid.n-apps\endcsname{24}
\expandafter\def\csname cnt@rev.desc.d6.uiv.p2.T.tdiag.center-d1.boot-resid.n-apps\endcsname{24}
\expandafter\def\csname pct@rev.desc.d6.uiv.p2.T.tdiag.center-d1.boot-resid.n-apps\endcsname{2400.0}
\expandafter\def\csname ppt@rev.desc.d6.uiv.p2.T.tdiag.center-d1.boot-resid.n-apps\endcsname{$+$2400.0}
\expandafter\def\csname val@rev.desc.d6.uiv.p2.T.tdiag.center-d1.boot-resid.n-boot-valid\endcsname{4000}
\expandafter\def\csname cnt@rev.desc.d6.uiv.p2.T.tdiag.center-d1.boot-resid.n-boot-valid\endcsname{4{,}000}
\expandafter\def\csname pct@rev.desc.d6.uiv.p2.T.tdiag.center-d1.boot-resid.n-boot-valid\endcsname{400000.0}
\expandafter\def\csname ppt@rev.desc.d6.uiv.p2.T.tdiag.center-d1.boot-resid.n-boot-valid\endcsname{$+$400000.0}
\expandafter\def\csname val@rev.desc.d6.uiv.p2.T.tdiag.center-d1.boot-resid.value\endcsname{-0.015}
\expandafter\def\csname cnt@rev.desc.d6.uiv.p2.T.tdiag.center-d1.boot-resid.value\endcsname{0}
\expandafter\def\csname pct@rev.desc.d6.uiv.p2.T.tdiag.center-d1.boot-resid.value\endcsname{$-$1.5}
\expandafter\def\csname ppt@rev.desc.d6.uiv.p2.T.tdiag.center-d1.boot-resid.value\endcsname{$-$1.5}
\expandafter\def\csname val@rev.desc.d6.uiv.p2.T.tdiag.center-d1.boot-share.ci95.0\endcsname{0.833}
\expandafter\def\csname cnt@rev.desc.d6.uiv.p2.T.tdiag.center-d1.boot-share.ci95.0\endcsname{1}
\expandafter\def\csname pct@rev.desc.d6.uiv.p2.T.tdiag.center-d1.boot-share.ci95.0\endcsname{83.3}
\expandafter\def\csname ppt@rev.desc.d6.uiv.p2.T.tdiag.center-d1.boot-share.ci95.0\endcsname{$+$83.3}
\expandafter\def\csname val@rev.desc.d6.uiv.p2.T.tdiag.center-d1.boot-share.ci95.1\endcsname{0.965}
\expandafter\def\csname cnt@rev.desc.d6.uiv.p2.T.tdiag.center-d1.boot-share.ci95.1\endcsname{1}
\expandafter\def\csname pct@rev.desc.d6.uiv.p2.T.tdiag.center-d1.boot-share.ci95.1\endcsname{96.5}
\expandafter\def\csname ppt@rev.desc.d6.uiv.p2.T.tdiag.center-d1.boot-share.ci95.1\endcsname{$+$96.5}
\expandafter\def\csname val@rev.desc.d6.uiv.p2.T.tdiag.center-d1.boot-share.n-apps\endcsname{24}
\expandafter\def\csname cnt@rev.desc.d6.uiv.p2.T.tdiag.center-d1.boot-share.n-apps\endcsname{24}
\expandafter\def\csname pct@rev.desc.d6.uiv.p2.T.tdiag.center-d1.boot-share.n-apps\endcsname{2400.0}
\expandafter\def\csname ppt@rev.desc.d6.uiv.p2.T.tdiag.center-d1.boot-share.n-apps\endcsname{$+$2400.0}
\expandafter\def\csname val@rev.desc.d6.uiv.p2.T.tdiag.center-d1.boot-share.n-boot-valid\endcsname{4000}
\expandafter\def\csname cnt@rev.desc.d6.uiv.p2.T.tdiag.center-d1.boot-share.n-boot-valid\endcsname{4{,}000}
\expandafter\def\csname pct@rev.desc.d6.uiv.p2.T.tdiag.center-d1.boot-share.n-boot-valid\endcsname{400000.0}
\expandafter\def\csname ppt@rev.desc.d6.uiv.p2.T.tdiag.center-d1.boot-share.n-boot-valid\endcsname{$+$400000.0}
\expandafter\def\csname val@rev.desc.d6.uiv.p2.T.tdiag.center-d1.boot-share.value\endcsname{0.903}
\expandafter\def\csname cnt@rev.desc.d6.uiv.p2.T.tdiag.center-d1.boot-share.value\endcsname{1}
\expandafter\def\csname pct@rev.desc.d6.uiv.p2.T.tdiag.center-d1.boot-share.value\endcsname{90.3}
\expandafter\def\csname ppt@rev.desc.d6.uiv.p2.T.tdiag.center-d1.boot-share.value\endcsname{$+$90.3}
\expandafter\def\csname val@rev.desc.d6.uiv.p2.T.tdiag.center-d1.n\endcsname{144}
\expandafter\def\csname cnt@rev.desc.d6.uiv.p2.T.tdiag.center-d1.n\endcsname{144}
\expandafter\def\csname pct@rev.desc.d6.uiv.p2.T.tdiag.center-d1.n\endcsname{14400.0}
\expandafter\def\csname ppt@rev.desc.d6.uiv.p2.T.tdiag.center-d1.n\endcsname{$+$14400.0}
\expandafter\def\csname val@rev.desc.d6.uiv.p2.T.tdiag.center-d1.resid-agree-stable\endcsname{-0.015}
\expandafter\def\csname cnt@rev.desc.d6.uiv.p2.T.tdiag.center-d1.resid-agree-stable\endcsname{0}
\expandafter\def\csname pct@rev.desc.d6.uiv.p2.T.tdiag.center-d1.resid-agree-stable\endcsname{$-$1.5}
\expandafter\def\csname ppt@rev.desc.d6.uiv.p2.T.tdiag.center-d1.resid-agree-stable\endcsname{$-$1.5}
\expandafter\def\csname val@rev.desc.d6.uiv.p2.T.tdiag.center-d1.resid-disagree\endcsname{0.111}
\expandafter\def\csname cnt@rev.desc.d6.uiv.p2.T.tdiag.center-d1.resid-disagree\endcsname{0}
\expandafter\def\csname pct@rev.desc.d6.uiv.p2.T.tdiag.center-d1.resid-disagree\endcsname{11.1}
\expandafter\def\csname ppt@rev.desc.d6.uiv.p2.T.tdiag.center-d1.resid-disagree\endcsname{$+$11.1}
\expandafter\def\csname val@rev.desc.d6.uiv.p2.T.tdiag.center-d1.share-agree-stable\endcsname{0.903}
\expandafter\def\csname cnt@rev.desc.d6.uiv.p2.T.tdiag.center-d1.share-agree-stable\endcsname{1}
\expandafter\def\csname pct@rev.desc.d6.uiv.p2.T.tdiag.center-d1.share-agree-stable\endcsname{90.3}
\expandafter\def\csname ppt@rev.desc.d6.uiv.p2.T.tdiag.center-d1.share-agree-stable\endcsname{$+$90.3}
\expandafter\def\csname val@rev.desc.d6.uiv.p2.T.tdiag.context-d1.N-agree-stable\endcsname{131}
\expandafter\def\csname cnt@rev.desc.d6.uiv.p2.T.tdiag.context-d1.N-agree-stable\endcsname{131}
\expandafter\def\csname pct@rev.desc.d6.uiv.p2.T.tdiag.context-d1.N-agree-stable\endcsname{13100.0}
\expandafter\def\csname ppt@rev.desc.d6.uiv.p2.T.tdiag.context-d1.N-agree-stable\endcsname{$+$13100.0}
\expandafter\def\csname val@rev.desc.d6.uiv.p2.T.tdiag.context-d1.N-disagree\endcsname{10}
\expandafter\def\csname cnt@rev.desc.d6.uiv.p2.T.tdiag.context-d1.N-disagree\endcsname{10}
\expandafter\def\csname pct@rev.desc.d6.uiv.p2.T.tdiag.context-d1.N-disagree\endcsname{1000.0}
\expandafter\def\csname ppt@rev.desc.d6.uiv.p2.T.tdiag.context-d1.N-disagree\endcsname{$+$1000.0}
\expandafter\def\csname val@rev.desc.d6.uiv.p2.T.tdiag.context-d1.boot-resid.ci95.0\endcsname{-0.023}
\expandafter\def\csname cnt@rev.desc.d6.uiv.p2.T.tdiag.context-d1.boot-resid.ci95.0\endcsname{0}
\expandafter\def\csname pct@rev.desc.d6.uiv.p2.T.tdiag.context-d1.boot-resid.ci95.0\endcsname{$-$2.3}
\expandafter\def\csname ppt@rev.desc.d6.uiv.p2.T.tdiag.context-d1.boot-resid.ci95.0\endcsname{$-$2.3}
\expandafter\def\csname val@rev.desc.d6.uiv.p2.T.tdiag.context-d1.boot-resid.ci95.1\endcsname{0.023}
\expandafter\def\csname cnt@rev.desc.d6.uiv.p2.T.tdiag.context-d1.boot-resid.ci95.1\endcsname{0}
\expandafter\def\csname pct@rev.desc.d6.uiv.p2.T.tdiag.context-d1.boot-resid.ci95.1\endcsname{2.3}
\expandafter\def\csname ppt@rev.desc.d6.uiv.p2.T.tdiag.context-d1.boot-resid.ci95.1\endcsname{$+$2.3}
\expandafter\def\csname val@rev.desc.d6.uiv.p2.T.tdiag.context-d1.boot-resid.n-apps\endcsname{24}
\expandafter\def\csname cnt@rev.desc.d6.uiv.p2.T.tdiag.context-d1.boot-resid.n-apps\endcsname{24}
\expandafter\def\csname pct@rev.desc.d6.uiv.p2.T.tdiag.context-d1.boot-resid.n-apps\endcsname{2400.0}
\expandafter\def\csname ppt@rev.desc.d6.uiv.p2.T.tdiag.context-d1.boot-resid.n-apps\endcsname{$+$2400.0}
\expandafter\def\csname val@rev.desc.d6.uiv.p2.T.tdiag.context-d1.boot-resid.n-boot-valid\endcsname{4000}
\expandafter\def\csname cnt@rev.desc.d6.uiv.p2.T.tdiag.context-d1.boot-resid.n-boot-valid\endcsname{4{,}000}
\expandafter\def\csname pct@rev.desc.d6.uiv.p2.T.tdiag.context-d1.boot-resid.n-boot-valid\endcsname{400000.0}
\expandafter\def\csname ppt@rev.desc.d6.uiv.p2.T.tdiag.context-d1.boot-resid.n-boot-valid\endcsname{$+$400000.0}
\expandafter\def\csname val@rev.desc.d6.uiv.p2.T.tdiag.context-d1.boot-resid.value\endcsname{0.000}
\expandafter\def\csname cnt@rev.desc.d6.uiv.p2.T.tdiag.context-d1.boot-resid.value\endcsname{0}
\expandafter\def\csname pct@rev.desc.d6.uiv.p2.T.tdiag.context-d1.boot-resid.value\endcsname{0.0}
\expandafter\def\csname ppt@rev.desc.d6.uiv.p2.T.tdiag.context-d1.boot-resid.value\endcsname{0.0}
\expandafter\def\csname val@rev.desc.d6.uiv.p2.T.tdiag.context-d1.boot-share.ci95.0\endcsname{0.854}
\expandafter\def\csname cnt@rev.desc.d6.uiv.p2.T.tdiag.context-d1.boot-share.ci95.0\endcsname{1}
\expandafter\def\csname pct@rev.desc.d6.uiv.p2.T.tdiag.context-d1.boot-share.ci95.0\endcsname{85.4}
\expandafter\def\csname ppt@rev.desc.d6.uiv.p2.T.tdiag.context-d1.boot-share.ci95.0\endcsname{$+$85.4}
\expandafter\def\csname val@rev.desc.d6.uiv.p2.T.tdiag.context-d1.boot-share.ci95.1\endcsname{0.958}
\expandafter\def\csname cnt@rev.desc.d6.uiv.p2.T.tdiag.context-d1.boot-share.ci95.1\endcsname{1}
\expandafter\def\csname pct@rev.desc.d6.uiv.p2.T.tdiag.context-d1.boot-share.ci95.1\endcsname{95.8}
\expandafter\def\csname ppt@rev.desc.d6.uiv.p2.T.tdiag.context-d1.boot-share.ci95.1\endcsname{$+$95.8}
\expandafter\def\csname val@rev.desc.d6.uiv.p2.T.tdiag.context-d1.boot-share.n-apps\endcsname{24}
\expandafter\def\csname cnt@rev.desc.d6.uiv.p2.T.tdiag.context-d1.boot-share.n-apps\endcsname{24}
\expandafter\def\csname pct@rev.desc.d6.uiv.p2.T.tdiag.context-d1.boot-share.n-apps\endcsname{2400.0}
\expandafter\def\csname ppt@rev.desc.d6.uiv.p2.T.tdiag.context-d1.boot-share.n-apps\endcsname{$+$2400.0}
\expandafter\def\csname val@rev.desc.d6.uiv.p2.T.tdiag.context-d1.boot-share.n-boot-valid\endcsname{4000}
\expandafter\def\csname cnt@rev.desc.d6.uiv.p2.T.tdiag.context-d1.boot-share.n-boot-valid\endcsname{4{,}000}
\expandafter\def\csname pct@rev.desc.d6.uiv.p2.T.tdiag.context-d1.boot-share.n-boot-valid\endcsname{400000.0}
\expandafter\def\csname ppt@rev.desc.d6.uiv.p2.T.tdiag.context-d1.boot-share.n-boot-valid\endcsname{$+$400000.0}
\expandafter\def\csname val@rev.desc.d6.uiv.p2.T.tdiag.context-d1.boot-share.value\endcsname{0.910}
\expandafter\def\csname cnt@rev.desc.d6.uiv.p2.T.tdiag.context-d1.boot-share.value\endcsname{1}
\expandafter\def\csname pct@rev.desc.d6.uiv.p2.T.tdiag.context-d1.boot-share.value\endcsname{91.0}
\expandafter\def\csname ppt@rev.desc.d6.uiv.p2.T.tdiag.context-d1.boot-share.value\endcsname{$+$91.0}
\expandafter\def\csname val@rev.desc.d6.uiv.p2.T.tdiag.context-d1.n\endcsname{144}
\expandafter\def\csname cnt@rev.desc.d6.uiv.p2.T.tdiag.context-d1.n\endcsname{144}
\expandafter\def\csname pct@rev.desc.d6.uiv.p2.T.tdiag.context-d1.n\endcsname{14400.0}
\expandafter\def\csname ppt@rev.desc.d6.uiv.p2.T.tdiag.context-d1.n\endcsname{$+$14400.0}
\expandafter\def\csname val@rev.desc.d6.uiv.p2.T.tdiag.context-d1.resid-agree-stable\endcsname{0.000}
\expandafter\def\csname cnt@rev.desc.d6.uiv.p2.T.tdiag.context-d1.resid-agree-stable\endcsname{0}
\expandafter\def\csname pct@rev.desc.d6.uiv.p2.T.tdiag.context-d1.resid-agree-stable\endcsname{0.0}
\expandafter\def\csname ppt@rev.desc.d6.uiv.p2.T.tdiag.context-d1.resid-agree-stable\endcsname{0.0}
\expandafter\def\csname val@rev.desc.d6.uiv.p2.T.tdiag.context-d1.resid-disagree\endcsname{0.400}
\expandafter\def\csname cnt@rev.desc.d6.uiv.p2.T.tdiag.context-d1.resid-disagree\endcsname{0}
\expandafter\def\csname pct@rev.desc.d6.uiv.p2.T.tdiag.context-d1.resid-disagree\endcsname{40.0}
\expandafter\def\csname ppt@rev.desc.d6.uiv.p2.T.tdiag.context-d1.resid-disagree\endcsname{$+$40.0}
\expandafter\def\csname val@rev.desc.d6.uiv.p2.T.tdiag.context-d1.share-agree-stable\endcsname{0.910}
\expandafter\def\csname cnt@rev.desc.d6.uiv.p2.T.tdiag.context-d1.share-agree-stable\endcsname{1}
\expandafter\def\csname pct@rev.desc.d6.uiv.p2.T.tdiag.context-d1.share-agree-stable\endcsname{91.0}
\expandafter\def\csname ppt@rev.desc.d6.uiv.p2.T.tdiag.context-d1.share-agree-stable\endcsname{$+$91.0}
\expandafter\def\csname val@rev.desc.d6.uiv.p2.T.tdiag.dual-d2.N-agree-stable\endcsname{130}
\expandafter\def\csname cnt@rev.desc.d6.uiv.p2.T.tdiag.dual-d2.N-agree-stable\endcsname{130}
\expandafter\def\csname pct@rev.desc.d6.uiv.p2.T.tdiag.dual-d2.N-agree-stable\endcsname{13000.0}
\expandafter\def\csname ppt@rev.desc.d6.uiv.p2.T.tdiag.dual-d2.N-agree-stable\endcsname{$+$13000.0}
\expandafter\def\csname val@rev.desc.d6.uiv.p2.T.tdiag.dual-d2.N-disagree\endcsname{9}
\expandafter\def\csname cnt@rev.desc.d6.uiv.p2.T.tdiag.dual-d2.N-disagree\endcsname{9}
\expandafter\def\csname pct@rev.desc.d6.uiv.p2.T.tdiag.dual-d2.N-disagree\endcsname{900.0}
\expandafter\def\csname ppt@rev.desc.d6.uiv.p2.T.tdiag.dual-d2.N-disagree\endcsname{$+$900.0}
\expandafter\def\csname val@rev.desc.d6.uiv.p2.T.tdiag.dual-d2.boot-resid.ci95.0\endcsname{-0.063}
\expandafter\def\csname cnt@rev.desc.d6.uiv.p2.T.tdiag.dual-d2.boot-resid.ci95.0\endcsname{0}
\expandafter\def\csname pct@rev.desc.d6.uiv.p2.T.tdiag.dual-d2.boot-resid.ci95.0\endcsname{$-$6.3}
\expandafter\def\csname ppt@rev.desc.d6.uiv.p2.T.tdiag.dual-d2.boot-resid.ci95.0\endcsname{$-$6.3}
\expandafter\def\csname val@rev.desc.d6.uiv.p2.T.tdiag.dual-d2.boot-resid.ci95.1\endcsname{0.031}
\expandafter\def\csname cnt@rev.desc.d6.uiv.p2.T.tdiag.dual-d2.boot-resid.ci95.1\endcsname{0}
\expandafter\def\csname pct@rev.desc.d6.uiv.p2.T.tdiag.dual-d2.boot-resid.ci95.1\endcsname{3.1}
\expandafter\def\csname ppt@rev.desc.d6.uiv.p2.T.tdiag.dual-d2.boot-resid.ci95.1\endcsname{$+$3.1}
\expandafter\def\csname val@rev.desc.d6.uiv.p2.T.tdiag.dual-d2.boot-resid.n-apps\endcsname{24}
\expandafter\def\csname cnt@rev.desc.d6.uiv.p2.T.tdiag.dual-d2.boot-resid.n-apps\endcsname{24}
\expandafter\def\csname pct@rev.desc.d6.uiv.p2.T.tdiag.dual-d2.boot-resid.n-apps\endcsname{2400.0}
\expandafter\def\csname ppt@rev.desc.d6.uiv.p2.T.tdiag.dual-d2.boot-resid.n-apps\endcsname{$+$2400.0}
\expandafter\def\csname val@rev.desc.d6.uiv.p2.T.tdiag.dual-d2.boot-resid.n-boot-valid\endcsname{4000}
\expandafter\def\csname cnt@rev.desc.d6.uiv.p2.T.tdiag.dual-d2.boot-resid.n-boot-valid\endcsname{4{,}000}
\expandafter\def\csname pct@rev.desc.d6.uiv.p2.T.tdiag.dual-d2.boot-resid.n-boot-valid\endcsname{400000.0}
\expandafter\def\csname ppt@rev.desc.d6.uiv.p2.T.tdiag.dual-d2.boot-resid.n-boot-valid\endcsname{$+$400000.0}
\expandafter\def\csname val@rev.desc.d6.uiv.p2.T.tdiag.dual-d2.boot-resid.value\endcsname{-0.015}
\expandafter\def\csname cnt@rev.desc.d6.uiv.p2.T.tdiag.dual-d2.boot-resid.value\endcsname{0}
\expandafter\def\csname pct@rev.desc.d6.uiv.p2.T.tdiag.dual-d2.boot-resid.value\endcsname{$-$1.5}
\expandafter\def\csname ppt@rev.desc.d6.uiv.p2.T.tdiag.dual-d2.boot-resid.value\endcsname{$-$1.5}
\expandafter\def\csname val@rev.desc.d6.uiv.p2.T.tdiag.dual-d2.boot-share.ci95.0\endcsname{0.847}
\expandafter\def\csname cnt@rev.desc.d6.uiv.p2.T.tdiag.dual-d2.boot-share.ci95.0\endcsname{1}
\expandafter\def\csname pct@rev.desc.d6.uiv.p2.T.tdiag.dual-d2.boot-share.ci95.0\endcsname{84.7}
\expandafter\def\csname ppt@rev.desc.d6.uiv.p2.T.tdiag.dual-d2.boot-share.ci95.0\endcsname{$+$84.7}
\expandafter\def\csname val@rev.desc.d6.uiv.p2.T.tdiag.dual-d2.boot-share.ci95.1\endcsname{0.951}
\expandafter\def\csname cnt@rev.desc.d6.uiv.p2.T.tdiag.dual-d2.boot-share.ci95.1\endcsname{1}
\expandafter\def\csname pct@rev.desc.d6.uiv.p2.T.tdiag.dual-d2.boot-share.ci95.1\endcsname{95.1}
\expandafter\def\csname ppt@rev.desc.d6.uiv.p2.T.tdiag.dual-d2.boot-share.ci95.1\endcsname{$+$95.1}
\expandafter\def\csname val@rev.desc.d6.uiv.p2.T.tdiag.dual-d2.boot-share.n-apps\endcsname{24}
\expandafter\def\csname cnt@rev.desc.d6.uiv.p2.T.tdiag.dual-d2.boot-share.n-apps\endcsname{24}
\expandafter\def\csname pct@rev.desc.d6.uiv.p2.T.tdiag.dual-d2.boot-share.n-apps\endcsname{2400.0}
\expandafter\def\csname ppt@rev.desc.d6.uiv.p2.T.tdiag.dual-d2.boot-share.n-apps\endcsname{$+$2400.0}
\expandafter\def\csname val@rev.desc.d6.uiv.p2.T.tdiag.dual-d2.boot-share.n-boot-valid\endcsname{4000}
\expandafter\def\csname cnt@rev.desc.d6.uiv.p2.T.tdiag.dual-d2.boot-share.n-boot-valid\endcsname{4{,}000}
\expandafter\def\csname pct@rev.desc.d6.uiv.p2.T.tdiag.dual-d2.boot-share.n-boot-valid\endcsname{400000.0}
\expandafter\def\csname ppt@rev.desc.d6.uiv.p2.T.tdiag.dual-d2.boot-share.n-boot-valid\endcsname{$+$400000.0}
\expandafter\def\csname val@rev.desc.d6.uiv.p2.T.tdiag.dual-d2.boot-share.value\endcsname{0.903}
\expandafter\def\csname cnt@rev.desc.d6.uiv.p2.T.tdiag.dual-d2.boot-share.value\endcsname{1}
\expandafter\def\csname pct@rev.desc.d6.uiv.p2.T.tdiag.dual-d2.boot-share.value\endcsname{90.3}
\expandafter\def\csname ppt@rev.desc.d6.uiv.p2.T.tdiag.dual-d2.boot-share.value\endcsname{$+$90.3}
\expandafter\def\csname val@rev.desc.d6.uiv.p2.T.tdiag.dual-d2.n\endcsname{144}
\expandafter\def\csname cnt@rev.desc.d6.uiv.p2.T.tdiag.dual-d2.n\endcsname{144}
\expandafter\def\csname pct@rev.desc.d6.uiv.p2.T.tdiag.dual-d2.n\endcsname{14400.0}
\expandafter\def\csname ppt@rev.desc.d6.uiv.p2.T.tdiag.dual-d2.n\endcsname{$+$14400.0}
\expandafter\def\csname val@rev.desc.d6.uiv.p2.T.tdiag.dual-d2.resid-agree-stable\endcsname{-0.015}
\expandafter\def\csname cnt@rev.desc.d6.uiv.p2.T.tdiag.dual-d2.resid-agree-stable\endcsname{0}
\expandafter\def\csname pct@rev.desc.d6.uiv.p2.T.tdiag.dual-d2.resid-agree-stable\endcsname{$-$1.5}
\expandafter\def\csname ppt@rev.desc.d6.uiv.p2.T.tdiag.dual-d2.resid-agree-stable\endcsname{$-$1.5}
\expandafter\def\csname val@rev.desc.d6.uiv.p2.T.tdiag.dual-d2.resid-disagree\endcsname{0.000}
\expandafter\def\csname cnt@rev.desc.d6.uiv.p2.T.tdiag.dual-d2.resid-disagree\endcsname{0}
\expandafter\def\csname pct@rev.desc.d6.uiv.p2.T.tdiag.dual-d2.resid-disagree\endcsname{0.0}
\expandafter\def\csname ppt@rev.desc.d6.uiv.p2.T.tdiag.dual-d2.resid-disagree\endcsname{0.0}
\expandafter\def\csname val@rev.desc.d6.uiv.p2.T.tdiag.dual-d2.share-agree-stable\endcsname{0.903}
\expandafter\def\csname cnt@rev.desc.d6.uiv.p2.T.tdiag.dual-d2.share-agree-stable\endcsname{1}
\expandafter\def\csname pct@rev.desc.d6.uiv.p2.T.tdiag.dual-d2.share-agree-stable\endcsname{90.3}
\expandafter\def\csname ppt@rev.desc.d6.uiv.p2.T.tdiag.dual-d2.share-agree-stable\endcsname{$+$90.3}
\expandafter\def\csname val@rev.desc.d6.uiv.p2.T.tdiag.grid2-d3.N-agree-stable\endcsname{125}
\expandafter\def\csname cnt@rev.desc.d6.uiv.p2.T.tdiag.grid2-d3.N-agree-stable\endcsname{125}
\expandafter\def\csname pct@rev.desc.d6.uiv.p2.T.tdiag.grid2-d3.N-agree-stable\endcsname{12500.0}
\expandafter\def\csname ppt@rev.desc.d6.uiv.p2.T.tdiag.grid2-d3.N-agree-stable\endcsname{$+$12500.0}
\expandafter\def\csname val@rev.desc.d6.uiv.p2.T.tdiag.grid2-d3.N-disagree\endcsname{15}
\expandafter\def\csname cnt@rev.desc.d6.uiv.p2.T.tdiag.grid2-d3.N-disagree\endcsname{15}
\expandafter\def\csname pct@rev.desc.d6.uiv.p2.T.tdiag.grid2-d3.N-disagree\endcsname{1500.0}
\expandafter\def\csname ppt@rev.desc.d6.uiv.p2.T.tdiag.grid2-d3.N-disagree\endcsname{$+$1500.0}
\expandafter\def\csname val@rev.desc.d6.uiv.p2.T.tdiag.grid2-d3.boot-resid.ci95.0\endcsname{-0.064}
\expandafter\def\csname cnt@rev.desc.d6.uiv.p2.T.tdiag.grid2-d3.boot-resid.ci95.0\endcsname{0}
\expandafter\def\csname pct@rev.desc.d6.uiv.p2.T.tdiag.grid2-d3.boot-resid.ci95.0\endcsname{$-$6.4}
\expandafter\def\csname ppt@rev.desc.d6.uiv.p2.T.tdiag.grid2-d3.boot-resid.ci95.0\endcsname{$-$6.4}
\expandafter\def\csname val@rev.desc.d6.uiv.p2.T.tdiag.grid2-d3.boot-resid.ci95.1\endcsname{0.000}
\expandafter\def\csname cnt@rev.desc.d6.uiv.p2.T.tdiag.grid2-d3.boot-resid.ci95.1\endcsname{0}
\expandafter\def\csname pct@rev.desc.d6.uiv.p2.T.tdiag.grid2-d3.boot-resid.ci95.1\endcsname{0.0}
\expandafter\def\csname ppt@rev.desc.d6.uiv.p2.T.tdiag.grid2-d3.boot-resid.ci95.1\endcsname{0.0}
\expandafter\def\csname val@rev.desc.d6.uiv.p2.T.tdiag.grid2-d3.boot-resid.n-apps\endcsname{24}
\expandafter\def\csname cnt@rev.desc.d6.uiv.p2.T.tdiag.grid2-d3.boot-resid.n-apps\endcsname{24}
\expandafter\def\csname pct@rev.desc.d6.uiv.p2.T.tdiag.grid2-d3.boot-resid.n-apps\endcsname{2400.0}
\expandafter\def\csname ppt@rev.desc.d6.uiv.p2.T.tdiag.grid2-d3.boot-resid.n-apps\endcsname{$+$2400.0}
\expandafter\def\csname val@rev.desc.d6.uiv.p2.T.tdiag.grid2-d3.boot-resid.n-boot-valid\endcsname{4000}
\expandafter\def\csname cnt@rev.desc.d6.uiv.p2.T.tdiag.grid2-d3.boot-resid.n-boot-valid\endcsname{4{,}000}
\expandafter\def\csname pct@rev.desc.d6.uiv.p2.T.tdiag.grid2-d3.boot-resid.n-boot-valid\endcsname{400000.0}
\expandafter\def\csname ppt@rev.desc.d6.uiv.p2.T.tdiag.grid2-d3.boot-resid.n-boot-valid\endcsname{$+$400000.0}
\expandafter\def\csname val@rev.desc.d6.uiv.p2.T.tdiag.grid2-d3.boot-resid.value\endcsname{-0.024}
\expandafter\def\csname cnt@rev.desc.d6.uiv.p2.T.tdiag.grid2-d3.boot-resid.value\endcsname{0}
\expandafter\def\csname pct@rev.desc.d6.uiv.p2.T.tdiag.grid2-d3.boot-resid.value\endcsname{$-$2.4}
\expandafter\def\csname ppt@rev.desc.d6.uiv.p2.T.tdiag.grid2-d3.boot-resid.value\endcsname{$-$2.4}
\expandafter\def\csname val@rev.desc.d6.uiv.p2.T.tdiag.grid2-d3.boot-share.ci95.0\endcsname{0.799}
\expandafter\def\csname cnt@rev.desc.d6.uiv.p2.T.tdiag.grid2-d3.boot-share.ci95.0\endcsname{1}
\expandafter\def\csname pct@rev.desc.d6.uiv.p2.T.tdiag.grid2-d3.boot-share.ci95.0\endcsname{79.9}
\expandafter\def\csname ppt@rev.desc.d6.uiv.p2.T.tdiag.grid2-d3.boot-share.ci95.0\endcsname{$+$79.9}
\expandafter\def\csname val@rev.desc.d6.uiv.p2.T.tdiag.grid2-d3.boot-share.ci95.1\endcsname{0.931}
\expandafter\def\csname cnt@rev.desc.d6.uiv.p2.T.tdiag.grid2-d3.boot-share.ci95.1\endcsname{1}
\expandafter\def\csname pct@rev.desc.d6.uiv.p2.T.tdiag.grid2-d3.boot-share.ci95.1\endcsname{93.1}
\expandafter\def\csname ppt@rev.desc.d6.uiv.p2.T.tdiag.grid2-d3.boot-share.ci95.1\endcsname{$+$93.1}
\expandafter\def\csname val@rev.desc.d6.uiv.p2.T.tdiag.grid2-d3.boot-share.n-apps\endcsname{24}
\expandafter\def\csname cnt@rev.desc.d6.uiv.p2.T.tdiag.grid2-d3.boot-share.n-apps\endcsname{24}
\expandafter\def\csname pct@rev.desc.d6.uiv.p2.T.tdiag.grid2-d3.boot-share.n-apps\endcsname{2400.0}
\expandafter\def\csname ppt@rev.desc.d6.uiv.p2.T.tdiag.grid2-d3.boot-share.n-apps\endcsname{$+$2400.0}
\expandafter\def\csname val@rev.desc.d6.uiv.p2.T.tdiag.grid2-d3.boot-share.n-boot-valid\endcsname{4000}
\expandafter\def\csname cnt@rev.desc.d6.uiv.p2.T.tdiag.grid2-d3.boot-share.n-boot-valid\endcsname{4{,}000}
\expandafter\def\csname pct@rev.desc.d6.uiv.p2.T.tdiag.grid2-d3.boot-share.n-boot-valid\endcsname{400000.0}
\expandafter\def\csname ppt@rev.desc.d6.uiv.p2.T.tdiag.grid2-d3.boot-share.n-boot-valid\endcsname{$+$400000.0}
\expandafter\def\csname val@rev.desc.d6.uiv.p2.T.tdiag.grid2-d3.boot-share.value\endcsname{0.868}
\expandafter\def\csname cnt@rev.desc.d6.uiv.p2.T.tdiag.grid2-d3.boot-share.value\endcsname{1}
\expandafter\def\csname pct@rev.desc.d6.uiv.p2.T.tdiag.grid2-d3.boot-share.value\endcsname{86.8}
\expandafter\def\csname ppt@rev.desc.d6.uiv.p2.T.tdiag.grid2-d3.boot-share.value\endcsname{$+$86.8}
\expandafter\def\csname val@rev.desc.d6.uiv.p2.T.tdiag.grid2-d3.n\endcsname{144}
\expandafter\def\csname cnt@rev.desc.d6.uiv.p2.T.tdiag.grid2-d3.n\endcsname{144}
\expandafter\def\csname pct@rev.desc.d6.uiv.p2.T.tdiag.grid2-d3.n\endcsname{14400.0}
\expandafter\def\csname ppt@rev.desc.d6.uiv.p2.T.tdiag.grid2-d3.n\endcsname{$+$14400.0}
\expandafter\def\csname val@rev.desc.d6.uiv.p2.T.tdiag.grid2-d3.resid-agree-stable\endcsname{-0.024}
\expandafter\def\csname cnt@rev.desc.d6.uiv.p2.T.tdiag.grid2-d3.resid-agree-stable\endcsname{0}
\expandafter\def\csname pct@rev.desc.d6.uiv.p2.T.tdiag.grid2-d3.resid-agree-stable\endcsname{$-$2.4}
\expandafter\def\csname ppt@rev.desc.d6.uiv.p2.T.tdiag.grid2-d3.resid-agree-stable\endcsname{$-$2.4}
\expandafter\def\csname val@rev.desc.d6.uiv.p2.T.tdiag.grid2-d3.resid-disagree\endcsname{0.467}
\expandafter\def\csname cnt@rev.desc.d6.uiv.p2.T.tdiag.grid2-d3.resid-disagree\endcsname{0}
\expandafter\def\csname pct@rev.desc.d6.uiv.p2.T.tdiag.grid2-d3.resid-disagree\endcsname{46.7}
\expandafter\def\csname ppt@rev.desc.d6.uiv.p2.T.tdiag.grid2-d3.resid-disagree\endcsname{$+$46.7}
\expandafter\def\csname val@rev.desc.d6.uiv.p2.T.tdiag.grid2-d3.share-agree-stable\endcsname{0.868}
\expandafter\def\csname cnt@rev.desc.d6.uiv.p2.T.tdiag.grid2-d3.share-agree-stable\endcsname{1}
\expandafter\def\csname pct@rev.desc.d6.uiv.p2.T.tdiag.grid2-d3.share-agree-stable\endcsname{86.8}
\expandafter\def\csname ppt@rev.desc.d6.uiv.p2.T.tdiag.grid2-d3.share-agree-stable\endcsname{$+$86.8}
\expandafter\def\csname val@rev.desc.d6.uiv.p2.T.tdiag.zoomclick-d3.N-agree-stable\endcsname{130}
\expandafter\def\csname cnt@rev.desc.d6.uiv.p2.T.tdiag.zoomclick-d3.N-agree-stable\endcsname{130}
\expandafter\def\csname pct@rev.desc.d6.uiv.p2.T.tdiag.zoomclick-d3.N-agree-stable\endcsname{13000.0}
\expandafter\def\csname ppt@rev.desc.d6.uiv.p2.T.tdiag.zoomclick-d3.N-agree-stable\endcsname{$+$13000.0}
\expandafter\def\csname val@rev.desc.d6.uiv.p2.T.tdiag.zoomclick-d3.N-disagree\endcsname{10}
\expandafter\def\csname cnt@rev.desc.d6.uiv.p2.T.tdiag.zoomclick-d3.N-disagree\endcsname{10}
\expandafter\def\csname pct@rev.desc.d6.uiv.p2.T.tdiag.zoomclick-d3.N-disagree\endcsname{1000.0}
\expandafter\def\csname ppt@rev.desc.d6.uiv.p2.T.tdiag.zoomclick-d3.N-disagree\endcsname{$+$1000.0}
\expandafter\def\csname val@rev.desc.d6.uiv.p2.T.tdiag.zoomclick-d3.boot-resid.ci95.0\endcsname{-0.016}
\expandafter\def\csname cnt@rev.desc.d6.uiv.p2.T.tdiag.zoomclick-d3.boot-resid.ci95.0\endcsname{0}
\expandafter\def\csname pct@rev.desc.d6.uiv.p2.T.tdiag.zoomclick-d3.boot-resid.ci95.0\endcsname{$-$1.6}
\expandafter\def\csname ppt@rev.desc.d6.uiv.p2.T.tdiag.zoomclick-d3.boot-resid.ci95.0\endcsname{$-$1.6}
\expandafter\def\csname val@rev.desc.d6.uiv.p2.T.tdiag.zoomclick-d3.boot-resid.ci95.1\endcsname{0.033}
\expandafter\def\csname cnt@rev.desc.d6.uiv.p2.T.tdiag.zoomclick-d3.boot-resid.ci95.1\endcsname{0}
\expandafter\def\csname pct@rev.desc.d6.uiv.p2.T.tdiag.zoomclick-d3.boot-resid.ci95.1\endcsname{3.3}
\expandafter\def\csname ppt@rev.desc.d6.uiv.p2.T.tdiag.zoomclick-d3.boot-resid.ci95.1\endcsname{$+$3.3}
\expandafter\def\csname val@rev.desc.d6.uiv.p2.T.tdiag.zoomclick-d3.boot-resid.n-apps\endcsname{24}
\expandafter\def\csname cnt@rev.desc.d6.uiv.p2.T.tdiag.zoomclick-d3.boot-resid.n-apps\endcsname{24}
\expandafter\def\csname pct@rev.desc.d6.uiv.p2.T.tdiag.zoomclick-d3.boot-resid.n-apps\endcsname{2400.0}
\expandafter\def\csname ppt@rev.desc.d6.uiv.p2.T.tdiag.zoomclick-d3.boot-resid.n-apps\endcsname{$+$2400.0}
\expandafter\def\csname val@rev.desc.d6.uiv.p2.T.tdiag.zoomclick-d3.boot-resid.n-boot-valid\endcsname{4000}
\expandafter\def\csname cnt@rev.desc.d6.uiv.p2.T.tdiag.zoomclick-d3.boot-resid.n-boot-valid\endcsname{4{,}000}
\expandafter\def\csname pct@rev.desc.d6.uiv.p2.T.tdiag.zoomclick-d3.boot-resid.n-boot-valid\endcsname{400000.0}
\expandafter\def\csname ppt@rev.desc.d6.uiv.p2.T.tdiag.zoomclick-d3.boot-resid.n-boot-valid\endcsname{$+$400000.0}
\expandafter\def\csname val@rev.desc.d6.uiv.p2.T.tdiag.zoomclick-d3.boot-resid.value\endcsname{0.008}
\expandafter\def\csname cnt@rev.desc.d6.uiv.p2.T.tdiag.zoomclick-d3.boot-resid.value\endcsname{0}
\expandafter\def\csname pct@rev.desc.d6.uiv.p2.T.tdiag.zoomclick-d3.boot-resid.value\endcsname{0.8}
\expandafter\def\csname ppt@rev.desc.d6.uiv.p2.T.tdiag.zoomclick-d3.boot-resid.value\endcsname{$+$0.8}
\expandafter\def\csname val@rev.desc.d6.uiv.p2.T.tdiag.zoomclick-d3.boot-share.ci95.0\endcsname{0.847}
\expandafter\def\csname cnt@rev.desc.d6.uiv.p2.T.tdiag.zoomclick-d3.boot-share.ci95.0\endcsname{1}
\expandafter\def\csname pct@rev.desc.d6.uiv.p2.T.tdiag.zoomclick-d3.boot-share.ci95.0\endcsname{84.7}
\expandafter\def\csname ppt@rev.desc.d6.uiv.p2.T.tdiag.zoomclick-d3.boot-share.ci95.0\endcsname{$+$84.7}
\expandafter\def\csname val@rev.desc.d6.uiv.p2.T.tdiag.zoomclick-d3.boot-share.ci95.1\endcsname{0.951}
\expandafter\def\csname cnt@rev.desc.d6.uiv.p2.T.tdiag.zoomclick-d3.boot-share.ci95.1\endcsname{1}
\expandafter\def\csname pct@rev.desc.d6.uiv.p2.T.tdiag.zoomclick-d3.boot-share.ci95.1\endcsname{95.1}
\expandafter\def\csname ppt@rev.desc.d6.uiv.p2.T.tdiag.zoomclick-d3.boot-share.ci95.1\endcsname{$+$95.1}
\expandafter\def\csname val@rev.desc.d6.uiv.p2.T.tdiag.zoomclick-d3.boot-share.n-apps\endcsname{24}
\expandafter\def\csname cnt@rev.desc.d6.uiv.p2.T.tdiag.zoomclick-d3.boot-share.n-apps\endcsname{24}
\expandafter\def\csname pct@rev.desc.d6.uiv.p2.T.tdiag.zoomclick-d3.boot-share.n-apps\endcsname{2400.0}
\expandafter\def\csname ppt@rev.desc.d6.uiv.p2.T.tdiag.zoomclick-d3.boot-share.n-apps\endcsname{$+$2400.0}
\expandafter\def\csname val@rev.desc.d6.uiv.p2.T.tdiag.zoomclick-d3.boot-share.n-boot-valid\endcsname{4000}
\expandafter\def\csname cnt@rev.desc.d6.uiv.p2.T.tdiag.zoomclick-d3.boot-share.n-boot-valid\endcsname{4{,}000}
\expandafter\def\csname pct@rev.desc.d6.uiv.p2.T.tdiag.zoomclick-d3.boot-share.n-boot-valid\endcsname{400000.0}
\expandafter\def\csname ppt@rev.desc.d6.uiv.p2.T.tdiag.zoomclick-d3.boot-share.n-boot-valid\endcsname{$+$400000.0}
\expandafter\def\csname val@rev.desc.d6.uiv.p2.T.tdiag.zoomclick-d3.boot-share.value\endcsname{0.903}
\expandafter\def\csname cnt@rev.desc.d6.uiv.p2.T.tdiag.zoomclick-d3.boot-share.value\endcsname{1}
\expandafter\def\csname pct@rev.desc.d6.uiv.p2.T.tdiag.zoomclick-d3.boot-share.value\endcsname{90.3}
\expandafter\def\csname ppt@rev.desc.d6.uiv.p2.T.tdiag.zoomclick-d3.boot-share.value\endcsname{$+$90.3}
\expandafter\def\csname val@rev.desc.d6.uiv.p2.T.tdiag.zoomclick-d3.n\endcsname{144}
\expandafter\def\csname cnt@rev.desc.d6.uiv.p2.T.tdiag.zoomclick-d3.n\endcsname{144}
\expandafter\def\csname pct@rev.desc.d6.uiv.p2.T.tdiag.zoomclick-d3.n\endcsname{14400.0}
\expandafter\def\csname ppt@rev.desc.d6.uiv.p2.T.tdiag.zoomclick-d3.n\endcsname{$+$14400.0}
\expandafter\def\csname val@rev.desc.d6.uiv.p2.T.tdiag.zoomclick-d3.resid-agree-stable\endcsname{0.008}
\expandafter\def\csname cnt@rev.desc.d6.uiv.p2.T.tdiag.zoomclick-d3.resid-agree-stable\endcsname{0}
\expandafter\def\csname pct@rev.desc.d6.uiv.p2.T.tdiag.zoomclick-d3.resid-agree-stable\endcsname{0.8}
\expandafter\def\csname ppt@rev.desc.d6.uiv.p2.T.tdiag.zoomclick-d3.resid-agree-stable\endcsname{$+$0.8}
\expandafter\def\csname val@rev.desc.d6.uiv.p2.T.tdiag.zoomclick-d3.resid-disagree\endcsname{0.100}
\expandafter\def\csname cnt@rev.desc.d6.uiv.p2.T.tdiag.zoomclick-d3.resid-disagree\endcsname{0}
\expandafter\def\csname pct@rev.desc.d6.uiv.p2.T.tdiag.zoomclick-d3.resid-disagree\endcsname{10.0}
\expandafter\def\csname ppt@rev.desc.d6.uiv.p2.T.tdiag.zoomclick-d3.resid-disagree\endcsname{$+$10.0}
\expandafter\def\csname val@rev.desc.d6.uiv.p2.T.tdiag.zoomclick-d3.share-agree-stable\endcsname{0.903}
\expandafter\def\csname cnt@rev.desc.d6.uiv.p2.T.tdiag.zoomclick-d3.share-agree-stable\endcsname{1}
\expandafter\def\csname pct@rev.desc.d6.uiv.p2.T.tdiag.zoomclick-d3.share-agree-stable\endcsname{90.3}
\expandafter\def\csname ppt@rev.desc.d6.uiv.p2.T.tdiag.zoomclick-d3.share-agree-stable\endcsname{$+$90.3}
\expandafter\def\csname val@rev.desc.d7.sspro.composition.A.category-n.OS\endcsname{16}
\expandafter\def\csname cnt@rev.desc.d7.sspro.composition.A.category-n.OS\endcsname{16}
\expandafter\def\csname pct@rev.desc.d7.sspro.composition.A.category-n.OS\endcsname{1600.0}
\expandafter\def\csname ppt@rev.desc.d7.sspro.composition.A.category-n.OS\endcsname{$+$1600.0}
\expandafter\def\csname val@rev.desc.d7.sspro.composition.A.category-n.Office\endcsname{8}
\expandafter\def\csname cnt@rev.desc.d7.sspro.composition.A.category-n.Office\endcsname{8}
\expandafter\def\csname pct@rev.desc.d7.sspro.composition.A.category-n.Office\endcsname{800.0}
\expandafter\def\csname ppt@rev.desc.d7.sspro.composition.A.category-n.Office\endcsname{$+$800.0}
\expandafter\def\csname val@rev.desc.d7.sspro.composition.A.category-n.Scientific\endcsname{8}
\expandafter\def\csname cnt@rev.desc.d7.sspro.composition.A.category-n.Scientific\endcsname{8}
\expandafter\def\csname pct@rev.desc.d7.sspro.composition.A.category-n.Scientific\endcsname{800.0}
\expandafter\def\csname ppt@rev.desc.d7.sspro.composition.A.category-n.Scientific\endcsname{$+$800.0}
\expandafter\def\csname val@rev.desc.d7.sspro.composition.A.n\endcsname{32}
\expandafter\def\csname cnt@rev.desc.d7.sspro.composition.A.n\endcsname{32}
\expandafter\def\csname pct@rev.desc.d7.sspro.composition.A.n\endcsname{3200.0}
\expandafter\def\csname ppt@rev.desc.d7.sspro.composition.A.n\endcsname{$+$3200.0}
\expandafter\def\csname val@rev.desc.d7.sspro.composition.A.n-icon\endcsname{16}
\expandafter\def\csname cnt@rev.desc.d7.sspro.composition.A.n-icon\endcsname{16}
\expandafter\def\csname pct@rev.desc.d7.sspro.composition.A.n-icon\endcsname{1600.0}
\expandafter\def\csname ppt@rev.desc.d7.sspro.composition.A.n-icon\endcsname{$+$1600.0}
\expandafter\def\csname val@rev.desc.d7.sspro.composition.A.n-text\endcsname{16}
\expandafter\def\csname cnt@rev.desc.d7.sspro.composition.A.n-text\endcsname{16}
\expandafter\def\csname pct@rev.desc.d7.sspro.composition.A.n-text\endcsname{1600.0}
\expandafter\def\csname ppt@rev.desc.d7.sspro.composition.A.n-text\endcsname{$+$1600.0}
\expandafter\def\csname val@rev.desc.d7.sspro.composition.A.share-icon\endcsname{0.500}
\expandafter\def\csname cnt@rev.desc.d7.sspro.composition.A.share-icon\endcsname{0}
\expandafter\def\csname pct@rev.desc.d7.sspro.composition.A.share-icon\endcsname{50.0}
\expandafter\def\csname ppt@rev.desc.d7.sspro.composition.A.share-icon\endcsname{$+$50.0}
\expandafter\def\csname val@rev.desc.d7.sspro.composition.A.share-text\endcsname{0.500}
\expandafter\def\csname cnt@rev.desc.d7.sspro.composition.A.share-text\endcsname{0}
\expandafter\def\csname pct@rev.desc.d7.sspro.composition.A.share-text\endcsname{50.0}
\expandafter\def\csname ppt@rev.desc.d7.sspro.composition.A.share-text\endcsname{$+$50.0}
\expandafter\def\csname val@rev.desc.d7.sspro.composition.C.category-n.Creative\endcsname{8}
\expandafter\def\csname cnt@rev.desc.d7.sspro.composition.C.category-n.Creative\endcsname{8}
\expandafter\def\csname pct@rev.desc.d7.sspro.composition.C.category-n.Creative\endcsname{800.0}
\expandafter\def\csname ppt@rev.desc.d7.sspro.composition.C.category-n.Creative\endcsname{$+$800.0}
\expandafter\def\csname val@rev.desc.d7.sspro.composition.C.category-n.Dev\endcsname{16}
\expandafter\def\csname cnt@rev.desc.d7.sspro.composition.C.category-n.Dev\endcsname{16}
\expandafter\def\csname pct@rev.desc.d7.sspro.composition.C.category-n.Dev\endcsname{1600.0}
\expandafter\def\csname ppt@rev.desc.d7.sspro.composition.C.category-n.Dev\endcsname{$+$1600.0}
\expandafter\def\csname val@rev.desc.d7.sspro.composition.C.category-n.OS\endcsname{8}
\expandafter\def\csname cnt@rev.desc.d7.sspro.composition.C.category-n.OS\endcsname{8}
\expandafter\def\csname pct@rev.desc.d7.sspro.composition.C.category-n.OS\endcsname{800.0}
\expandafter\def\csname ppt@rev.desc.d7.sspro.composition.C.category-n.OS\endcsname{$+$800.0}
\expandafter\def\csname val@rev.desc.d7.sspro.composition.C.category-n.Office\endcsname{8}
\expandafter\def\csname cnt@rev.desc.d7.sspro.composition.C.category-n.Office\endcsname{8}
\expandafter\def\csname pct@rev.desc.d7.sspro.composition.C.category-n.Office\endcsname{800.0}
\expandafter\def\csname ppt@rev.desc.d7.sspro.composition.C.category-n.Office\endcsname{$+$800.0}
\expandafter\def\csname val@rev.desc.d7.sspro.composition.C.category-n.Scientific\endcsname{8}
\expandafter\def\csname cnt@rev.desc.d7.sspro.composition.C.category-n.Scientific\endcsname{8}
\expandafter\def\csname pct@rev.desc.d7.sspro.composition.C.category-n.Scientific\endcsname{800.0}
\expandafter\def\csname ppt@rev.desc.d7.sspro.composition.C.category-n.Scientific\endcsname{$+$800.0}
\expandafter\def\csname val@rev.desc.d7.sspro.composition.C.n\endcsname{48}
\expandafter\def\csname cnt@rev.desc.d7.sspro.composition.C.n\endcsname{48}
\expandafter\def\csname pct@rev.desc.d7.sspro.composition.C.n\endcsname{4800.0}
\expandafter\def\csname ppt@rev.desc.d7.sspro.composition.C.n\endcsname{$+$4800.0}
\expandafter\def\csname val@rev.desc.d7.sspro.composition.C.n-icon\endcsname{26}
\expandafter\def\csname cnt@rev.desc.d7.sspro.composition.C.n-icon\endcsname{26}
\expandafter\def\csname pct@rev.desc.d7.sspro.composition.C.n-icon\endcsname{2600.0}
\expandafter\def\csname ppt@rev.desc.d7.sspro.composition.C.n-icon\endcsname{$+$2600.0}
\expandafter\def\csname val@rev.desc.d7.sspro.composition.C.n-text\endcsname{22}
\expandafter\def\csname cnt@rev.desc.d7.sspro.composition.C.n-text\endcsname{22}
\expandafter\def\csname pct@rev.desc.d7.sspro.composition.C.n-text\endcsname{2200.0}
\expandafter\def\csname ppt@rev.desc.d7.sspro.composition.C.n-text\endcsname{$+$2200.0}
\expandafter\def\csname val@rev.desc.d7.sspro.composition.C.share-icon\endcsname{0.542}
\expandafter\def\csname cnt@rev.desc.d7.sspro.composition.C.share-icon\endcsname{1}
\expandafter\def\csname pct@rev.desc.d7.sspro.composition.C.share-icon\endcsname{54.2}
\expandafter\def\csname ppt@rev.desc.d7.sspro.composition.C.share-icon\endcsname{$+$54.2}
\expandafter\def\csname val@rev.desc.d7.sspro.composition.C.share-text\endcsname{0.458}
\expandafter\def\csname cnt@rev.desc.d7.sspro.composition.C.share-text\endcsname{0}
\expandafter\def\csname pct@rev.desc.d7.sspro.composition.C.share-text\endcsname{45.8}
\expandafter\def\csname ppt@rev.desc.d7.sspro.composition.C.share-text\endcsname{$+$45.8}
\expandafter\def\csname val@rev.desc.d7.sspro.composition.T.category-n.CAD\endcsname{24}
\expandafter\def\csname cnt@rev.desc.d7.sspro.composition.T.category-n.CAD\endcsname{24}
\expandafter\def\csname pct@rev.desc.d7.sspro.composition.T.category-n.CAD\endcsname{2400.0}
\expandafter\def\csname ppt@rev.desc.d7.sspro.composition.T.category-n.CAD\endcsname{$+$2400.0}
\expandafter\def\csname val@rev.desc.d7.sspro.composition.T.category-n.Creative\endcsname{32}
\expandafter\def\csname cnt@rev.desc.d7.sspro.composition.T.category-n.Creative\endcsname{32}
\expandafter\def\csname pct@rev.desc.d7.sspro.composition.T.category-n.Creative\endcsname{3200.0}
\expandafter\def\csname ppt@rev.desc.d7.sspro.composition.T.category-n.Creative\endcsname{$+$3200.0}
\expandafter\def\csname val@rev.desc.d7.sspro.composition.T.category-n.Dev\endcsname{24}
\expandafter\def\csname cnt@rev.desc.d7.sspro.composition.T.category-n.Dev\endcsname{24}
\expandafter\def\csname pct@rev.desc.d7.sspro.composition.T.category-n.Dev\endcsname{2400.0}
\expandafter\def\csname ppt@rev.desc.d7.sspro.composition.T.category-n.Dev\endcsname{$+$2400.0}
\expandafter\def\csname val@rev.desc.d7.sspro.composition.T.category-n.Office\endcsname{8}
\expandafter\def\csname cnt@rev.desc.d7.sspro.composition.T.category-n.Office\endcsname{8}
\expandafter\def\csname pct@rev.desc.d7.sspro.composition.T.category-n.Office\endcsname{800.0}
\expandafter\def\csname ppt@rev.desc.d7.sspro.composition.T.category-n.Office\endcsname{$+$800.0}
\expandafter\def\csname val@rev.desc.d7.sspro.composition.T.category-n.Scientific\endcsname{8}
\expandafter\def\csname cnt@rev.desc.d7.sspro.composition.T.category-n.Scientific\endcsname{8}
\expandafter\def\csname pct@rev.desc.d7.sspro.composition.T.category-n.Scientific\endcsname{800.0}
\expandafter\def\csname ppt@rev.desc.d7.sspro.composition.T.category-n.Scientific\endcsname{$+$800.0}
\expandafter\def\csname val@rev.desc.d7.sspro.composition.T.n\endcsname{96}
\expandafter\def\csname cnt@rev.desc.d7.sspro.composition.T.n\endcsname{96}
\expandafter\def\csname pct@rev.desc.d7.sspro.composition.T.n\endcsname{9600.0}
\expandafter\def\csname ppt@rev.desc.d7.sspro.composition.T.n\endcsname{$+$9600.0}
\expandafter\def\csname val@rev.desc.d7.sspro.composition.T.n-icon\endcsname{28}
\expandafter\def\csname cnt@rev.desc.d7.sspro.composition.T.n-icon\endcsname{28}
\expandafter\def\csname pct@rev.desc.d7.sspro.composition.T.n-icon\endcsname{2800.0}
\expandafter\def\csname ppt@rev.desc.d7.sspro.composition.T.n-icon\endcsname{$+$2800.0}
\expandafter\def\csname val@rev.desc.d7.sspro.composition.T.n-text\endcsname{68}
\expandafter\def\csname cnt@rev.desc.d7.sspro.composition.T.n-text\endcsname{68}
\expandafter\def\csname pct@rev.desc.d7.sspro.composition.T.n-text\endcsname{6800.0}
\expandafter\def\csname ppt@rev.desc.d7.sspro.composition.T.n-text\endcsname{$+$6800.0}
\expandafter\def\csname val@rev.desc.d7.sspro.composition.T.share-icon\endcsname{0.292}
\expandafter\def\csname cnt@rev.desc.d7.sspro.composition.T.share-icon\endcsname{0}
\expandafter\def\csname pct@rev.desc.d7.sspro.composition.T.share-icon\endcsname{29.2}
\expandafter\def\csname ppt@rev.desc.d7.sspro.composition.T.share-icon\endcsname{$+$29.2}
\expandafter\def\csname val@rev.desc.d7.sspro.composition.T.share-text\endcsname{0.708}
\expandafter\def\csname cnt@rev.desc.d7.sspro.composition.T.share-text\endcsname{1}
\expandafter\def\csname pct@rev.desc.d7.sspro.composition.T.share-text\endcsname{70.8}
\expandafter\def\csname ppt@rev.desc.d7.sspro.composition.T.share-text\endcsname{$+$70.8}
\expandafter\def\csname val@rev.desc.d7.sspro.tau-T-by-type.p1.center-d1.icon-minus-text-on-apps-with-both.ci90.0\endcsname{-0.060}
\expandafter\def\csname cnt@rev.desc.d7.sspro.tau-T-by-type.p1.center-d1.icon-minus-text-on-apps-with-both.ci90.0\endcsname{0}
\expandafter\def\csname pct@rev.desc.d7.sspro.tau-T-by-type.p1.center-d1.icon-minus-text-on-apps-with-both.ci90.0\endcsname{$-$6.0}
\expandafter\def\csname ppt@rev.desc.d7.sspro.tau-T-by-type.p1.center-d1.icon-minus-text-on-apps-with-both.ci90.0\endcsname{$-$6.0}
\expandafter\def\csname val@rev.desc.d7.sspro.tau-T-by-type.p1.center-d1.icon-minus-text-on-apps-with-both.ci90.1\endcsname{0.340}
\expandafter\def\csname cnt@rev.desc.d7.sspro.tau-T-by-type.p1.center-d1.icon-minus-text-on-apps-with-both.ci90.1\endcsname{0}
\expandafter\def\csname pct@rev.desc.d7.sspro.tau-T-by-type.p1.center-d1.icon-minus-text-on-apps-with-both.ci90.1\endcsname{34.0}
\expandafter\def\csname ppt@rev.desc.d7.sspro.tau-T-by-type.p1.center-d1.icon-minus-text-on-apps-with-both.ci90.1\endcsname{$+$34.0}
\expandafter\def\csname val@rev.desc.d7.sspro.tau-T-by-type.p1.center-d1.icon-minus-text-on-apps-with-both.ci95-excludes-0\endcsname{0}
\expandafter\def\csname cnt@rev.desc.d7.sspro.tau-T-by-type.p1.center-d1.icon-minus-text-on-apps-with-both.ci95-excludes-0\endcsname{0}
\expandafter\def\csname pct@rev.desc.d7.sspro.tau-T-by-type.p1.center-d1.icon-minus-text-on-apps-with-both.ci95-excludes-0\endcsname{0.0}
\expandafter\def\csname ppt@rev.desc.d7.sspro.tau-T-by-type.p1.center-d1.icon-minus-text-on-apps-with-both.ci95-excludes-0\endcsname{0.0}
\expandafter\def\csname val@rev.desc.d7.sspro.tau-T-by-type.p1.center-d1.icon-minus-text-on-apps-with-both.ci95.0\endcsname{-0.097}
\expandafter\def\csname cnt@rev.desc.d7.sspro.tau-T-by-type.p1.center-d1.icon-minus-text-on-apps-with-both.ci95.0\endcsname{0}
\expandafter\def\csname pct@rev.desc.d7.sspro.tau-T-by-type.p1.center-d1.icon-minus-text-on-apps-with-both.ci95.0\endcsname{$-$9.7}
\expandafter\def\csname ppt@rev.desc.d7.sspro.tau-T-by-type.p1.center-d1.icon-minus-text-on-apps-with-both.ci95.0\endcsname{$-$9.7}
\expandafter\def\csname val@rev.desc.d7.sspro.tau-T-by-type.p1.center-d1.icon-minus-text-on-apps-with-both.ci95.1\endcsname{0.383}
\expandafter\def\csname cnt@rev.desc.d7.sspro.tau-T-by-type.p1.center-d1.icon-minus-text-on-apps-with-both.ci95.1\endcsname{0}
\expandafter\def\csname pct@rev.desc.d7.sspro.tau-T-by-type.p1.center-d1.icon-minus-text-on-apps-with-both.ci95.1\endcsname{38.3}
\expandafter\def\csname ppt@rev.desc.d7.sspro.tau-T-by-type.p1.center-d1.icon-minus-text-on-apps-with-both.ci95.1\endcsname{$+$38.3}
\expandafter\def\csname val@rev.desc.d7.sspro.tau-T-by-type.p1.center-d1.icon-minus-text-on-apps-with-both.mean\endcsname{0.130}
\expandafter\def\csname cnt@rev.desc.d7.sspro.tau-T-by-type.p1.center-d1.icon-minus-text-on-apps-with-both.mean\endcsname{0}
\expandafter\def\csname pct@rev.desc.d7.sspro.tau-T-by-type.p1.center-d1.icon-minus-text-on-apps-with-both.mean\endcsname{13.0}
\expandafter\def\csname ppt@rev.desc.d7.sspro.tau-T-by-type.p1.center-d1.icon-minus-text-on-apps-with-both.mean\endcsname{$+$13.0}
\expandafter\def\csname val@rev.desc.d7.sspro.tau-T-by-type.p1.center-d1.icon-minus-text-on-apps-with-both.n-apps\endcsname{10}
\expandafter\def\csname cnt@rev.desc.d7.sspro.tau-T-by-type.p1.center-d1.icon-minus-text-on-apps-with-both.n-apps\endcsname{10}
\expandafter\def\csname pct@rev.desc.d7.sspro.tau-T-by-type.p1.center-d1.icon-minus-text-on-apps-with-both.n-apps\endcsname{1000.0}
\expandafter\def\csname ppt@rev.desc.d7.sspro.tau-T-by-type.p1.center-d1.icon-minus-text-on-apps-with-both.n-apps\endcsname{$+$1000.0}
\expandafter\def\csname val@rev.desc.d7.sspro.tau-T-by-type.p1.center-d1.icon-minus-text-on-apps-with-both.tost1.ci90-inside\endcsname{0}
\expandafter\def\csname cnt@rev.desc.d7.sspro.tau-T-by-type.p1.center-d1.icon-minus-text-on-apps-with-both.tost1.ci90-inside\endcsname{0}
\expandafter\def\csname pct@rev.desc.d7.sspro.tau-T-by-type.p1.center-d1.icon-minus-text-on-apps-with-both.tost1.ci90-inside\endcsname{0.0}
\expandafter\def\csname ppt@rev.desc.d7.sspro.tau-T-by-type.p1.center-d1.icon-minus-text-on-apps-with-both.tost1.ci90-inside\endcsname{0.0}
\expandafter\def\csname val@rev.desc.d7.sspro.tau-T-by-type.p1.center-d1.icon-minus-text-on-apps-with-both.tost1.equivalent\endcsname{0}
\expandafter\def\csname cnt@rev.desc.d7.sspro.tau-T-by-type.p1.center-d1.icon-minus-text-on-apps-with-both.tost1.equivalent\endcsname{0}
\expandafter\def\csname pct@rev.desc.d7.sspro.tau-T-by-type.p1.center-d1.icon-minus-text-on-apps-with-both.tost1.equivalent\endcsname{0.0}
\expandafter\def\csname ppt@rev.desc.d7.sspro.tau-T-by-type.p1.center-d1.icon-minus-text-on-apps-with-both.tost1.equivalent\endcsname{0.0}
\expandafter\def\csname val@rev.desc.d7.sspro.tau-T-by-type.p1.center-d1.icon-minus-text-on-apps-with-both.tost1.p\endcsname{0.849}
\expandafter\def\csname cnt@rev.desc.d7.sspro.tau-T-by-type.p1.center-d1.icon-minus-text-on-apps-with-both.tost1.p\endcsname{1}
\expandafter\def\csname pct@rev.desc.d7.sspro.tau-T-by-type.p1.center-d1.icon-minus-text-on-apps-with-both.tost1.p\endcsname{84.9}
\expandafter\def\csname ppt@rev.desc.d7.sspro.tau-T-by-type.p1.center-d1.icon-minus-text-on-apps-with-both.tost1.p\endcsname{$+$84.9}
\expandafter\def\csname val@rev.desc.d7.sspro.tau-T-by-type.p1.center-d1.icon-minus-text-on-apps-with-both.tost5.ci90-inside\endcsname{0}
\expandafter\def\csname cnt@rev.desc.d7.sspro.tau-T-by-type.p1.center-d1.icon-minus-text-on-apps-with-both.tost5.ci90-inside\endcsname{0}
\expandafter\def\csname pct@rev.desc.d7.sspro.tau-T-by-type.p1.center-d1.icon-minus-text-on-apps-with-both.tost5.ci90-inside\endcsname{0.0}
\expandafter\def\csname ppt@rev.desc.d7.sspro.tau-T-by-type.p1.center-d1.icon-minus-text-on-apps-with-both.tost5.ci90-inside\endcsname{0.0}
\expandafter\def\csname val@rev.desc.d7.sspro.tau-T-by-type.p1.center-d1.icon-minus-text-on-apps-with-both.tost5.equivalent\endcsname{0}
\expandafter\def\csname cnt@rev.desc.d7.sspro.tau-T-by-type.p1.center-d1.icon-minus-text-on-apps-with-both.tost5.equivalent\endcsname{0}
\expandafter\def\csname pct@rev.desc.d7.sspro.tau-T-by-type.p1.center-d1.icon-minus-text-on-apps-with-both.tost5.equivalent\endcsname{0.0}
\expandafter\def\csname ppt@rev.desc.d7.sspro.tau-T-by-type.p1.center-d1.icon-minus-text-on-apps-with-both.tost5.equivalent\endcsname{0.0}
\expandafter\def\csname val@rev.desc.d7.sspro.tau-T-by-type.p1.center-d1.icon-minus-text-on-apps-with-both.tost5.p\endcsname{0.743}
\expandafter\def\csname cnt@rev.desc.d7.sspro.tau-T-by-type.p1.center-d1.icon-minus-text-on-apps-with-both.tost5.p\endcsname{1}
\expandafter\def\csname pct@rev.desc.d7.sspro.tau-T-by-type.p1.center-d1.icon-minus-text-on-apps-with-both.tost5.p\endcsname{74.3}
\expandafter\def\csname ppt@rev.desc.d7.sspro.tau-T-by-type.p1.center-d1.icon-minus-text-on-apps-with-both.tost5.p\endcsname{$+$74.3}
\expandafter\def\csname val@rev.desc.d7.sspro.tau-T-by-type.p1.center-d1.icon.acc-F\endcsname{0.714}
\expandafter\def\csname cnt@rev.desc.d7.sspro.tau-T-by-type.p1.center-d1.icon.acc-F\endcsname{1}
\expandafter\def\csname pct@rev.desc.d7.sspro.tau-T-by-type.p1.center-d1.icon.acc-F\endcsname{71.4}
\expandafter\def\csname ppt@rev.desc.d7.sspro.tau-T-by-type.p1.center-d1.icon.acc-F\endcsname{$+$71.4}
\expandafter\def\csname val@rev.desc.d7.sspro.tau-T-by-type.p1.center-d1.icon.acc-direct\endcsname{0.643}
\expandafter\def\csname cnt@rev.desc.d7.sspro.tau-T-by-type.p1.center-d1.icon.acc-direct\endcsname{1}
\expandafter\def\csname pct@rev.desc.d7.sspro.tau-T-by-type.p1.center-d1.icon.acc-direct\endcsname{64.3}
\expandafter\def\csname ppt@rev.desc.d7.sspro.tau-T-by-type.p1.center-d1.icon.acc-direct\endcsname{$+$64.3}
\expandafter\def\csname val@rev.desc.d7.sspro.tau-T-by-type.p1.center-d1.icon.n-apps\endcsname{10}
\expandafter\def\csname cnt@rev.desc.d7.sspro.tau-T-by-type.p1.center-d1.icon.n-apps\endcsname{10}
\expandafter\def\csname pct@rev.desc.d7.sspro.tau-T-by-type.p1.center-d1.icon.n-apps\endcsname{1000.0}
\expandafter\def\csname ppt@rev.desc.d7.sspro.tau-T-by-type.p1.center-d1.icon.n-apps\endcsname{$+$1000.0}
\expandafter\def\csname val@rev.desc.d7.sspro.tau-T-by-type.p1.center-d1.icon.n-tasks\endcsname{28}
\expandafter\def\csname cnt@rev.desc.d7.sspro.tau-T-by-type.p1.center-d1.icon.n-tasks\endcsname{28}
\expandafter\def\csname pct@rev.desc.d7.sspro.tau-T-by-type.p1.center-d1.icon.n-tasks\endcsname{2800.0}
\expandafter\def\csname ppt@rev.desc.d7.sspro.tau-T-by-type.p1.center-d1.icon.n-tasks\endcsname{$+$2800.0}
\expandafter\def\csname val@rev.desc.d7.sspro.tau-T-by-type.p1.center-d1.icon.tau.ci90.0\endcsname{-0.020}
\expandafter\def\csname cnt@rev.desc.d7.sspro.tau-T-by-type.p1.center-d1.icon.tau.ci90.0\endcsname{0}
\expandafter\def\csname pct@rev.desc.d7.sspro.tau-T-by-type.p1.center-d1.icon.tau.ci90.0\endcsname{$-$2.0}
\expandafter\def\csname ppt@rev.desc.d7.sspro.tau-T-by-type.p1.center-d1.icon.tau.ci90.0\endcsname{$-$2.0}
\expandafter\def\csname val@rev.desc.d7.sspro.tau-T-by-type.p1.center-d1.icon.tau.ci90.1\endcsname{0.370}
\expandafter\def\csname cnt@rev.desc.d7.sspro.tau-T-by-type.p1.center-d1.icon.tau.ci90.1\endcsname{0}
\expandafter\def\csname pct@rev.desc.d7.sspro.tau-T-by-type.p1.center-d1.icon.tau.ci90.1\endcsname{37.0}
\expandafter\def\csname ppt@rev.desc.d7.sspro.tau-T-by-type.p1.center-d1.icon.tau.ci90.1\endcsname{$+$37.0}
\expandafter\def\csname val@rev.desc.d7.sspro.tau-T-by-type.p1.center-d1.icon.tau.ci95-excludes-0\endcsname{0}
\expandafter\def\csname cnt@rev.desc.d7.sspro.tau-T-by-type.p1.center-d1.icon.tau.ci95-excludes-0\endcsname{0}
\expandafter\def\csname pct@rev.desc.d7.sspro.tau-T-by-type.p1.center-d1.icon.tau.ci95-excludes-0\endcsname{0.0}
\expandafter\def\csname ppt@rev.desc.d7.sspro.tau-T-by-type.p1.center-d1.icon.tau.ci95-excludes-0\endcsname{0.0}
\expandafter\def\csname val@rev.desc.d7.sspro.tau-T-by-type.p1.center-d1.icon.tau.ci95.0\endcsname{-0.050}
\expandafter\def\csname cnt@rev.desc.d7.sspro.tau-T-by-type.p1.center-d1.icon.tau.ci95.0\endcsname{0}
\expandafter\def\csname pct@rev.desc.d7.sspro.tau-T-by-type.p1.center-d1.icon.tau.ci95.0\endcsname{$-$5.0}
\expandafter\def\csname ppt@rev.desc.d7.sspro.tau-T-by-type.p1.center-d1.icon.tau.ci95.0\endcsname{$-$5.0}
\expandafter\def\csname val@rev.desc.d7.sspro.tau-T-by-type.p1.center-d1.icon.tau.ci95.1\endcsname{0.410}
\expandafter\def\csname cnt@rev.desc.d7.sspro.tau-T-by-type.p1.center-d1.icon.tau.ci95.1\endcsname{0}
\expandafter\def\csname pct@rev.desc.d7.sspro.tau-T-by-type.p1.center-d1.icon.tau.ci95.1\endcsname{41.0}
\expandafter\def\csname ppt@rev.desc.d7.sspro.tau-T-by-type.p1.center-d1.icon.tau.ci95.1\endcsname{$+$41.0}
\expandafter\def\csname val@rev.desc.d7.sspro.tau-T-by-type.p1.center-d1.icon.tau.mean\endcsname{0.167}
\expandafter\def\csname cnt@rev.desc.d7.sspro.tau-T-by-type.p1.center-d1.icon.tau.mean\endcsname{0}
\expandafter\def\csname pct@rev.desc.d7.sspro.tau-T-by-type.p1.center-d1.icon.tau.mean\endcsname{16.7}
\expandafter\def\csname ppt@rev.desc.d7.sspro.tau-T-by-type.p1.center-d1.icon.tau.mean\endcsname{$+$16.7}
\expandafter\def\csname val@rev.desc.d7.sspro.tau-T-by-type.p1.center-d1.icon.tau.n-apps\endcsname{10}
\expandafter\def\csname cnt@rev.desc.d7.sspro.tau-T-by-type.p1.center-d1.icon.tau.n-apps\endcsname{10}
\expandafter\def\csname pct@rev.desc.d7.sspro.tau-T-by-type.p1.center-d1.icon.tau.n-apps\endcsname{1000.0}
\expandafter\def\csname ppt@rev.desc.d7.sspro.tau-T-by-type.p1.center-d1.icon.tau.n-apps\endcsname{$+$1000.0}
\expandafter\def\csname val@rev.desc.d7.sspro.tau-T-by-type.p1.center-d1.icon.tau.tost1.ci90-inside\endcsname{0}
\expandafter\def\csname cnt@rev.desc.d7.sspro.tau-T-by-type.p1.center-d1.icon.tau.tost1.ci90-inside\endcsname{0}
\expandafter\def\csname pct@rev.desc.d7.sspro.tau-T-by-type.p1.center-d1.icon.tau.tost1.ci90-inside\endcsname{0.0}
\expandafter\def\csname ppt@rev.desc.d7.sspro.tau-T-by-type.p1.center-d1.icon.tau.tost1.ci90-inside\endcsname{0.0}
\expandafter\def\csname val@rev.desc.d7.sspro.tau-T-by-type.p1.center-d1.icon.tau.tost1.equivalent\endcsname{0}
\expandafter\def\csname cnt@rev.desc.d7.sspro.tau-T-by-type.p1.center-d1.icon.tau.tost1.equivalent\endcsname{0}
\expandafter\def\csname pct@rev.desc.d7.sspro.tau-T-by-type.p1.center-d1.icon.tau.tost1.equivalent\endcsname{0.0}
\expandafter\def\csname ppt@rev.desc.d7.sspro.tau-T-by-type.p1.center-d1.icon.tau.tost1.equivalent\endcsname{0.0}
\expandafter\def\csname val@rev.desc.d7.sspro.tau-T-by-type.p1.center-d1.icon.tau.tost1.p\endcsname{0.911}
\expandafter\def\csname cnt@rev.desc.d7.sspro.tau-T-by-type.p1.center-d1.icon.tau.tost1.p\endcsname{1}
\expandafter\def\csname pct@rev.desc.d7.sspro.tau-T-by-type.p1.center-d1.icon.tau.tost1.p\endcsname{91.1}
\expandafter\def\csname ppt@rev.desc.d7.sspro.tau-T-by-type.p1.center-d1.icon.tau.tost1.p\endcsname{$+$91.1}
\expandafter\def\csname val@rev.desc.d7.sspro.tau-T-by-type.p1.center-d1.icon.tau.tost5.ci90-inside\endcsname{0}
\expandafter\def\csname cnt@rev.desc.d7.sspro.tau-T-by-type.p1.center-d1.icon.tau.tost5.ci90-inside\endcsname{0}
\expandafter\def\csname pct@rev.desc.d7.sspro.tau-T-by-type.p1.center-d1.icon.tau.tost5.ci90-inside\endcsname{0.0}
\expandafter\def\csname ppt@rev.desc.d7.sspro.tau-T-by-type.p1.center-d1.icon.tau.tost5.ci90-inside\endcsname{0.0}
\expandafter\def\csname val@rev.desc.d7.sspro.tau-T-by-type.p1.center-d1.icon.tau.tost5.equivalent\endcsname{0}
\expandafter\def\csname cnt@rev.desc.d7.sspro.tau-T-by-type.p1.center-d1.icon.tau.tost5.equivalent\endcsname{0}
\expandafter\def\csname pct@rev.desc.d7.sspro.tau-T-by-type.p1.center-d1.icon.tau.tost5.equivalent\endcsname{0.0}
\expandafter\def\csname ppt@rev.desc.d7.sspro.tau-T-by-type.p1.center-d1.icon.tau.tost5.equivalent\endcsname{0.0}
\expandafter\def\csname val@rev.desc.d7.sspro.tau-T-by-type.p1.center-d1.icon.tau.tost5.p\endcsname{0.829}
\expandafter\def\csname cnt@rev.desc.d7.sspro.tau-T-by-type.p1.center-d1.icon.tau.tost5.p\endcsname{1}
\expandafter\def\csname pct@rev.desc.d7.sspro.tau-T-by-type.p1.center-d1.icon.tau.tost5.p\endcsname{82.9}
\expandafter\def\csname ppt@rev.desc.d7.sspro.tau-T-by-type.p1.center-d1.icon.tau.tost5.p\endcsname{$+$82.9}
\expandafter\def\csname val@rev.desc.d7.sspro.tau-T-by-type.p1.center-d1.text.acc-F\endcsname{0.897}
\expandafter\def\csname cnt@rev.desc.d7.sspro.tau-T-by-type.p1.center-d1.text.acc-F\endcsname{1}
\expandafter\def\csname pct@rev.desc.d7.sspro.tau-T-by-type.p1.center-d1.text.acc-F\endcsname{89.7}
\expandafter\def\csname ppt@rev.desc.d7.sspro.tau-T-by-type.p1.center-d1.text.acc-F\endcsname{$+$89.7}
\expandafter\def\csname val@rev.desc.d7.sspro.tau-T-by-type.p1.center-d1.text.acc-direct\endcsname{0.779}
\expandafter\def\csname cnt@rev.desc.d7.sspro.tau-T-by-type.p1.center-d1.text.acc-direct\endcsname{1}
\expandafter\def\csname pct@rev.desc.d7.sspro.tau-T-by-type.p1.center-d1.text.acc-direct\endcsname{77.9}
\expandafter\def\csname ppt@rev.desc.d7.sspro.tau-T-by-type.p1.center-d1.text.acc-direct\endcsname{$+$77.9}
\expandafter\def\csname val@rev.desc.d7.sspro.tau-T-by-type.p1.center-d1.text.n-apps\endcsname{12}
\expandafter\def\csname cnt@rev.desc.d7.sspro.tau-T-by-type.p1.center-d1.text.n-apps\endcsname{12}
\expandafter\def\csname pct@rev.desc.d7.sspro.tau-T-by-type.p1.center-d1.text.n-apps\endcsname{1200.0}
\expandafter\def\csname ppt@rev.desc.d7.sspro.tau-T-by-type.p1.center-d1.text.n-apps\endcsname{$+$1200.0}
\expandafter\def\csname val@rev.desc.d7.sspro.tau-T-by-type.p1.center-d1.text.n-tasks\endcsname{68}
\expandafter\def\csname cnt@rev.desc.d7.sspro.tau-T-by-type.p1.center-d1.text.n-tasks\endcsname{68}
\expandafter\def\csname pct@rev.desc.d7.sspro.tau-T-by-type.p1.center-d1.text.n-tasks\endcsname{6800.0}
\expandafter\def\csname ppt@rev.desc.d7.sspro.tau-T-by-type.p1.center-d1.text.n-tasks\endcsname{$+$6800.0}
\expandafter\def\csname val@rev.desc.d7.sspro.tau-T-by-type.p1.center-d1.text.tau.ci90.0\endcsname{0.021}
\expandafter\def\csname cnt@rev.desc.d7.sspro.tau-T-by-type.p1.center-d1.text.tau.ci90.0\endcsname{0}
\expandafter\def\csname pct@rev.desc.d7.sspro.tau-T-by-type.p1.center-d1.text.tau.ci90.0\endcsname{2.1}
\expandafter\def\csname ppt@rev.desc.d7.sspro.tau-T-by-type.p1.center-d1.text.tau.ci90.0\endcsname{$+$2.1}
\expandafter\def\csname val@rev.desc.d7.sspro.tau-T-by-type.p1.center-d1.text.tau.ci90.1\endcsname{0.184}
\expandafter\def\csname cnt@rev.desc.d7.sspro.tau-T-by-type.p1.center-d1.text.tau.ci90.1\endcsname{0}
\expandafter\def\csname pct@rev.desc.d7.sspro.tau-T-by-type.p1.center-d1.text.tau.ci90.1\endcsname{18.4}
\expandafter\def\csname ppt@rev.desc.d7.sspro.tau-T-by-type.p1.center-d1.text.tau.ci90.1\endcsname{$+$18.4}
\expandafter\def\csname val@rev.desc.d7.sspro.tau-T-by-type.p1.center-d1.text.tau.ci95-excludes-0\endcsname{1}
\expandafter\def\csname cnt@rev.desc.d7.sspro.tau-T-by-type.p1.center-d1.text.tau.ci95-excludes-0\endcsname{1}
\expandafter\def\csname pct@rev.desc.d7.sspro.tau-T-by-type.p1.center-d1.text.tau.ci95-excludes-0\endcsname{100.0}
\expandafter\def\csname ppt@rev.desc.d7.sspro.tau-T-by-type.p1.center-d1.text.tau.ci95-excludes-0\endcsname{$+$100.0}
\expandafter\def\csname val@rev.desc.d7.sspro.tau-T-by-type.p1.center-d1.text.tau.ci95.0\endcsname{0.014}
\expandafter\def\csname cnt@rev.desc.d7.sspro.tau-T-by-type.p1.center-d1.text.tau.ci95.0\endcsname{0}
\expandafter\def\csname pct@rev.desc.d7.sspro.tau-T-by-type.p1.center-d1.text.tau.ci95.0\endcsname{1.4}
\expandafter\def\csname ppt@rev.desc.d7.sspro.tau-T-by-type.p1.center-d1.text.tau.ci95.0\endcsname{$+$1.4}
\expandafter\def\csname val@rev.desc.d7.sspro.tau-T-by-type.p1.center-d1.text.tau.ci95.1\endcsname{0.205}
\expandafter\def\csname cnt@rev.desc.d7.sspro.tau-T-by-type.p1.center-d1.text.tau.ci95.1\endcsname{0}
\expandafter\def\csname pct@rev.desc.d7.sspro.tau-T-by-type.p1.center-d1.text.tau.ci95.1\endcsname{20.5}
\expandafter\def\csname ppt@rev.desc.d7.sspro.tau-T-by-type.p1.center-d1.text.tau.ci95.1\endcsname{$+$20.5}
\expandafter\def\csname val@rev.desc.d7.sspro.tau-T-by-type.p1.center-d1.text.tau.mean\endcsname{0.093}
\expandafter\def\csname cnt@rev.desc.d7.sspro.tau-T-by-type.p1.center-d1.text.tau.mean\endcsname{0}
\expandafter\def\csname pct@rev.desc.d7.sspro.tau-T-by-type.p1.center-d1.text.tau.mean\endcsname{9.3}
\expandafter\def\csname ppt@rev.desc.d7.sspro.tau-T-by-type.p1.center-d1.text.tau.mean\endcsname{$+$9.3}
\expandafter\def\csname val@rev.desc.d7.sspro.tau-T-by-type.p1.center-d1.text.tau.n-apps\endcsname{12}
\expandafter\def\csname cnt@rev.desc.d7.sspro.tau-T-by-type.p1.center-d1.text.tau.n-apps\endcsname{12}
\expandafter\def\csname pct@rev.desc.d7.sspro.tau-T-by-type.p1.center-d1.text.tau.n-apps\endcsname{1200.0}
\expandafter\def\csname ppt@rev.desc.d7.sspro.tau-T-by-type.p1.center-d1.text.tau.n-apps\endcsname{$+$1200.0}
\expandafter\def\csname val@rev.desc.d7.sspro.tau-T-by-type.p1.center-d1.text.tau.tost1.ci90-inside\endcsname{0}
\expandafter\def\csname cnt@rev.desc.d7.sspro.tau-T-by-type.p1.center-d1.text.tau.tost1.ci90-inside\endcsname{0}
\expandafter\def\csname pct@rev.desc.d7.sspro.tau-T-by-type.p1.center-d1.text.tau.tost1.ci90-inside\endcsname{0.0}
\expandafter\def\csname ppt@rev.desc.d7.sspro.tau-T-by-type.p1.center-d1.text.tau.tost1.ci90-inside\endcsname{0.0}
\expandafter\def\csname val@rev.desc.d7.sspro.tau-T-by-type.p1.center-d1.text.tau.tost1.equivalent\endcsname{0}
\expandafter\def\csname cnt@rev.desc.d7.sspro.tau-T-by-type.p1.center-d1.text.tau.tost1.equivalent\endcsname{0}
\expandafter\def\csname pct@rev.desc.d7.sspro.tau-T-by-type.p1.center-d1.text.tau.tost1.equivalent\endcsname{0.0}
\expandafter\def\csname ppt@rev.desc.d7.sspro.tau-T-by-type.p1.center-d1.text.tau.tost1.equivalent\endcsname{0.0}
\expandafter\def\csname val@rev.desc.d7.sspro.tau-T-by-type.p1.center-d1.text.tau.tost1.p\endcsname{0.990}
\expandafter\def\csname cnt@rev.desc.d7.sspro.tau-T-by-type.p1.center-d1.text.tau.tost1.p\endcsname{1}
\expandafter\def\csname pct@rev.desc.d7.sspro.tau-T-by-type.p1.center-d1.text.tau.tost1.p\endcsname{99.0}
\expandafter\def\csname ppt@rev.desc.d7.sspro.tau-T-by-type.p1.center-d1.text.tau.tost1.p\endcsname{$+$99.0}
\expandafter\def\csname val@rev.desc.d7.sspro.tau-T-by-type.p1.center-d1.text.tau.tost5.ci90-inside\endcsname{0}
\expandafter\def\csname cnt@rev.desc.d7.sspro.tau-T-by-type.p1.center-d1.text.tau.tost5.ci90-inside\endcsname{0}
\expandafter\def\csname pct@rev.desc.d7.sspro.tau-T-by-type.p1.center-d1.text.tau.tost5.ci90-inside\endcsname{0.0}
\expandafter\def\csname ppt@rev.desc.d7.sspro.tau-T-by-type.p1.center-d1.text.tau.tost5.ci90-inside\endcsname{0.0}
\expandafter\def\csname val@rev.desc.d7.sspro.tau-T-by-type.p1.center-d1.text.tau.tost5.equivalent\endcsname{0}
\expandafter\def\csname cnt@rev.desc.d7.sspro.tau-T-by-type.p1.center-d1.text.tau.tost5.equivalent\endcsname{0}
\expandafter\def\csname pct@rev.desc.d7.sspro.tau-T-by-type.p1.center-d1.text.tau.tost5.equivalent\endcsname{0.0}
\expandafter\def\csname ppt@rev.desc.d7.sspro.tau-T-by-type.p1.center-d1.text.tau.tost5.equivalent\endcsname{0.0}
\expandafter\def\csname val@rev.desc.d7.sspro.tau-T-by-type.p1.center-d1.text.tau.tost5.p\endcsname{0.789}
\expandafter\def\csname cnt@rev.desc.d7.sspro.tau-T-by-type.p1.center-d1.text.tau.tost5.p\endcsname{1}
\expandafter\def\csname pct@rev.desc.d7.sspro.tau-T-by-type.p1.center-d1.text.tau.tost5.p\endcsname{78.9}
\expandafter\def\csname ppt@rev.desc.d7.sspro.tau-T-by-type.p1.center-d1.text.tau.tost5.p\endcsname{$+$78.9}
\expandafter\def\csname val@rev.desc.d7.sspro.tau-T-by-type.p1.context-d1.icon-minus-text-on-apps-with-both.ci90.0\endcsname{-0.030}
\expandafter\def\csname cnt@rev.desc.d7.sspro.tau-T-by-type.p1.context-d1.icon-minus-text-on-apps-with-both.ci90.0\endcsname{0}
\expandafter\def\csname pct@rev.desc.d7.sspro.tau-T-by-type.p1.context-d1.icon-minus-text-on-apps-with-both.ci90.0\endcsname{$-$3.0}
\expandafter\def\csname ppt@rev.desc.d7.sspro.tau-T-by-type.p1.context-d1.icon-minus-text-on-apps-with-both.ci90.0\endcsname{$-$3.0}
\expandafter\def\csname val@rev.desc.d7.sspro.tau-T-by-type.p1.context-d1.icon-minus-text-on-apps-with-both.ci90.1\endcsname{0.442}
\expandafter\def\csname cnt@rev.desc.d7.sspro.tau-T-by-type.p1.context-d1.icon-minus-text-on-apps-with-both.ci90.1\endcsname{0}
\expandafter\def\csname pct@rev.desc.d7.sspro.tau-T-by-type.p1.context-d1.icon-minus-text-on-apps-with-both.ci90.1\endcsname{44.2}
\expandafter\def\csname ppt@rev.desc.d7.sspro.tau-T-by-type.p1.context-d1.icon-minus-text-on-apps-with-both.ci90.1\endcsname{$+$44.2}
\expandafter\def\csname val@rev.desc.d7.sspro.tau-T-by-type.p1.context-d1.icon-minus-text-on-apps-with-both.ci95-excludes-0\endcsname{0}
\expandafter\def\csname cnt@rev.desc.d7.sspro.tau-T-by-type.p1.context-d1.icon-minus-text-on-apps-with-both.ci95-excludes-0\endcsname{0}
\expandafter\def\csname pct@rev.desc.d7.sspro.tau-T-by-type.p1.context-d1.icon-minus-text-on-apps-with-both.ci95-excludes-0\endcsname{0.0}
\expandafter\def\csname ppt@rev.desc.d7.sspro.tau-T-by-type.p1.context-d1.icon-minus-text-on-apps-with-both.ci95-excludes-0\endcsname{0.0}
\expandafter\def\csname val@rev.desc.d7.sspro.tau-T-by-type.p1.context-d1.icon-minus-text-on-apps-with-both.ci95.0\endcsname{-0.074}
\expandafter\def\csname cnt@rev.desc.d7.sspro.tau-T-by-type.p1.context-d1.icon-minus-text-on-apps-with-both.ci95.0\endcsname{0}
\expandafter\def\csname pct@rev.desc.d7.sspro.tau-T-by-type.p1.context-d1.icon-minus-text-on-apps-with-both.ci95.0\endcsname{$-$7.4}
\expandafter\def\csname ppt@rev.desc.d7.sspro.tau-T-by-type.p1.context-d1.icon-minus-text-on-apps-with-both.ci95.0\endcsname{$-$7.4}
\expandafter\def\csname val@rev.desc.d7.sspro.tau-T-by-type.p1.context-d1.icon-minus-text-on-apps-with-both.ci95.1\endcsname{0.486}
\expandafter\def\csname cnt@rev.desc.d7.sspro.tau-T-by-type.p1.context-d1.icon-minus-text-on-apps-with-both.ci95.1\endcsname{0}
\expandafter\def\csname pct@rev.desc.d7.sspro.tau-T-by-type.p1.context-d1.icon-minus-text-on-apps-with-both.ci95.1\endcsname{48.6}
\expandafter\def\csname ppt@rev.desc.d7.sspro.tau-T-by-type.p1.context-d1.icon-minus-text-on-apps-with-both.ci95.1\endcsname{$+$48.6}
\expandafter\def\csname val@rev.desc.d7.sspro.tau-T-by-type.p1.context-d1.icon-minus-text-on-apps-with-both.mean\endcsname{0.197}
\expandafter\def\csname cnt@rev.desc.d7.sspro.tau-T-by-type.p1.context-d1.icon-minus-text-on-apps-with-both.mean\endcsname{0}
\expandafter\def\csname pct@rev.desc.d7.sspro.tau-T-by-type.p1.context-d1.icon-minus-text-on-apps-with-both.mean\endcsname{19.7}
\expandafter\def\csname ppt@rev.desc.d7.sspro.tau-T-by-type.p1.context-d1.icon-minus-text-on-apps-with-both.mean\endcsname{$+$19.7}
\expandafter\def\csname val@rev.desc.d7.sspro.tau-T-by-type.p1.context-d1.icon-minus-text-on-apps-with-both.n-apps\endcsname{10}
\expandafter\def\csname cnt@rev.desc.d7.sspro.tau-T-by-type.p1.context-d1.icon-minus-text-on-apps-with-both.n-apps\endcsname{10}
\expandafter\def\csname pct@rev.desc.d7.sspro.tau-T-by-type.p1.context-d1.icon-minus-text-on-apps-with-both.n-apps\endcsname{1000.0}
\expandafter\def\csname ppt@rev.desc.d7.sspro.tau-T-by-type.p1.context-d1.icon-minus-text-on-apps-with-both.n-apps\endcsname{$+$1000.0}
\expandafter\def\csname val@rev.desc.d7.sspro.tau-T-by-type.p1.context-d1.icon-minus-text-on-apps-with-both.tost1.ci90-inside\endcsname{0}
\expandafter\def\csname cnt@rev.desc.d7.sspro.tau-T-by-type.p1.context-d1.icon-minus-text-on-apps-with-both.tost1.ci90-inside\endcsname{0}
\expandafter\def\csname pct@rev.desc.d7.sspro.tau-T-by-type.p1.context-d1.icon-minus-text-on-apps-with-both.tost1.ci90-inside\endcsname{0.0}
\expandafter\def\csname ppt@rev.desc.d7.sspro.tau-T-by-type.p1.context-d1.icon-minus-text-on-apps-with-both.tost1.ci90-inside\endcsname{0.0}
\expandafter\def\csname val@rev.desc.d7.sspro.tau-T-by-type.p1.context-d1.icon-minus-text-on-apps-with-both.tost1.equivalent\endcsname{0}
\expandafter\def\csname cnt@rev.desc.d7.sspro.tau-T-by-type.p1.context-d1.icon-minus-text-on-apps-with-both.tost1.equivalent\endcsname{0}
\expandafter\def\csname pct@rev.desc.d7.sspro.tau-T-by-type.p1.context-d1.icon-minus-text-on-apps-with-both.tost1.equivalent\endcsname{0.0}
\expandafter\def\csname ppt@rev.desc.d7.sspro.tau-T-by-type.p1.context-d1.icon-minus-text-on-apps-with-both.tost1.equivalent\endcsname{0.0}
\expandafter\def\csname val@rev.desc.d7.sspro.tau-T-by-type.p1.context-d1.icon-minus-text-on-apps-with-both.tost1.p\endcsname{0.906}
\expandafter\def\csname cnt@rev.desc.d7.sspro.tau-T-by-type.p1.context-d1.icon-minus-text-on-apps-with-both.tost1.p\endcsname{1}
\expandafter\def\csname pct@rev.desc.d7.sspro.tau-T-by-type.p1.context-d1.icon-minus-text-on-apps-with-both.tost1.p\endcsname{90.6}
\expandafter\def\csname ppt@rev.desc.d7.sspro.tau-T-by-type.p1.context-d1.icon-minus-text-on-apps-with-both.tost1.p\endcsname{$+$90.6}
\expandafter\def\csname val@rev.desc.d7.sspro.tau-T-by-type.p1.context-d1.icon-minus-text-on-apps-with-both.tost5.ci90-inside\endcsname{0}
\expandafter\def\csname cnt@rev.desc.d7.sspro.tau-T-by-type.p1.context-d1.icon-minus-text-on-apps-with-both.tost5.ci90-inside\endcsname{0}
\expandafter\def\csname pct@rev.desc.d7.sspro.tau-T-by-type.p1.context-d1.icon-minus-text-on-apps-with-both.tost5.ci90-inside\endcsname{0.0}
\expandafter\def\csname ppt@rev.desc.d7.sspro.tau-T-by-type.p1.context-d1.icon-minus-text-on-apps-with-both.tost5.ci90-inside\endcsname{0.0}
\expandafter\def\csname val@rev.desc.d7.sspro.tau-T-by-type.p1.context-d1.icon-minus-text-on-apps-with-both.tost5.equivalent\endcsname{0}
\expandafter\def\csname cnt@rev.desc.d7.sspro.tau-T-by-type.p1.context-d1.icon-minus-text-on-apps-with-both.tost5.equivalent\endcsname{0}
\expandafter\def\csname pct@rev.desc.d7.sspro.tau-T-by-type.p1.context-d1.icon-minus-text-on-apps-with-both.tost5.equivalent\endcsname{0.0}
\expandafter\def\csname ppt@rev.desc.d7.sspro.tau-T-by-type.p1.context-d1.icon-minus-text-on-apps-with-both.tost5.equivalent\endcsname{0.0}
\expandafter\def\csname val@rev.desc.d7.sspro.tau-T-by-type.p1.context-d1.icon-minus-text-on-apps-with-both.tost5.p\endcsname{0.849}
\expandafter\def\csname cnt@rev.desc.d7.sspro.tau-T-by-type.p1.context-d1.icon-minus-text-on-apps-with-both.tost5.p\endcsname{1}
\expandafter\def\csname pct@rev.desc.d7.sspro.tau-T-by-type.p1.context-d1.icon-minus-text-on-apps-with-both.tost5.p\endcsname{84.9}
\expandafter\def\csname ppt@rev.desc.d7.sspro.tau-T-by-type.p1.context-d1.icon-minus-text-on-apps-with-both.tost5.p\endcsname{$+$84.9}
\expandafter\def\csname val@rev.desc.d7.sspro.tau-T-by-type.p1.context-d1.icon.acc-F\endcsname{0.750}
\expandafter\def\csname cnt@rev.desc.d7.sspro.tau-T-by-type.p1.context-d1.icon.acc-F\endcsname{1}
\expandafter\def\csname pct@rev.desc.d7.sspro.tau-T-by-type.p1.context-d1.icon.acc-F\endcsname{75.0}
\expandafter\def\csname ppt@rev.desc.d7.sspro.tau-T-by-type.p1.context-d1.icon.acc-F\endcsname{$+$75.0}
\expandafter\def\csname val@rev.desc.d7.sspro.tau-T-by-type.p1.context-d1.icon.acc-direct\endcsname{0.643}
\expandafter\def\csname cnt@rev.desc.d7.sspro.tau-T-by-type.p1.context-d1.icon.acc-direct\endcsname{1}
\expandafter\def\csname pct@rev.desc.d7.sspro.tau-T-by-type.p1.context-d1.icon.acc-direct\endcsname{64.3}
\expandafter\def\csname ppt@rev.desc.d7.sspro.tau-T-by-type.p1.context-d1.icon.acc-direct\endcsname{$+$64.3}
\expandafter\def\csname val@rev.desc.d7.sspro.tau-T-by-type.p1.context-d1.icon.n-apps\endcsname{10}
\expandafter\def\csname cnt@rev.desc.d7.sspro.tau-T-by-type.p1.context-d1.icon.n-apps\endcsname{10}
\expandafter\def\csname pct@rev.desc.d7.sspro.tau-T-by-type.p1.context-d1.icon.n-apps\endcsname{1000.0}
\expandafter\def\csname ppt@rev.desc.d7.sspro.tau-T-by-type.p1.context-d1.icon.n-apps\endcsname{$+$1000.0}
\expandafter\def\csname val@rev.desc.d7.sspro.tau-T-by-type.p1.context-d1.icon.n-tasks\endcsname{28}
\expandafter\def\csname cnt@rev.desc.d7.sspro.tau-T-by-type.p1.context-d1.icon.n-tasks\endcsname{28}
\expandafter\def\csname pct@rev.desc.d7.sspro.tau-T-by-type.p1.context-d1.icon.n-tasks\endcsname{2800.0}
\expandafter\def\csname ppt@rev.desc.d7.sspro.tau-T-by-type.p1.context-d1.icon.n-tasks\endcsname{$+$2800.0}
\expandafter\def\csname val@rev.desc.d7.sspro.tau-T-by-type.p1.context-d1.icon.tau.ci90.0\endcsname{0.010}
\expandafter\def\csname cnt@rev.desc.d7.sspro.tau-T-by-type.p1.context-d1.icon.tau.ci90.0\endcsname{0}
\expandafter\def\csname pct@rev.desc.d7.sspro.tau-T-by-type.p1.context-d1.icon.tau.ci90.0\endcsname{1.0}
\expandafter\def\csname ppt@rev.desc.d7.sspro.tau-T-by-type.p1.context-d1.icon.tau.ci90.0\endcsname{$+$1.0}
\expandafter\def\csname val@rev.desc.d7.sspro.tau-T-by-type.p1.context-d1.icon.tau.ci90.1\endcsname{0.403}
\expandafter\def\csname cnt@rev.desc.d7.sspro.tau-T-by-type.p1.context-d1.icon.tau.ci90.1\endcsname{0}
\expandafter\def\csname pct@rev.desc.d7.sspro.tau-T-by-type.p1.context-d1.icon.tau.ci90.1\endcsname{40.3}
\expandafter\def\csname ppt@rev.desc.d7.sspro.tau-T-by-type.p1.context-d1.icon.tau.ci90.1\endcsname{$+$40.3}
\expandafter\def\csname val@rev.desc.d7.sspro.tau-T-by-type.p1.context-d1.icon.tau.ci95-excludes-0\endcsname{0}
\expandafter\def\csname cnt@rev.desc.d7.sspro.tau-T-by-type.p1.context-d1.icon.tau.ci95-excludes-0\endcsname{0}
\expandafter\def\csname pct@rev.desc.d7.sspro.tau-T-by-type.p1.context-d1.icon.tau.ci95-excludes-0\endcsname{0.0}
\expandafter\def\csname ppt@rev.desc.d7.sspro.tau-T-by-type.p1.context-d1.icon.tau.ci95-excludes-0\endcsname{0.0}
\expandafter\def\csname val@rev.desc.d7.sspro.tau-T-by-type.p1.context-d1.icon.tau.ci95.0\endcsname{-0.017}
\expandafter\def\csname cnt@rev.desc.d7.sspro.tau-T-by-type.p1.context-d1.icon.tau.ci95.0\endcsname{0}
\expandafter\def\csname pct@rev.desc.d7.sspro.tau-T-by-type.p1.context-d1.icon.tau.ci95.0\endcsname{$-$1.7}
\expandafter\def\csname ppt@rev.desc.d7.sspro.tau-T-by-type.p1.context-d1.icon.tau.ci95.0\endcsname{$-$1.7}
\expandafter\def\csname val@rev.desc.d7.sspro.tau-T-by-type.p1.context-d1.icon.tau.ci95.1\endcsname{0.450}
\expandafter\def\csname cnt@rev.desc.d7.sspro.tau-T-by-type.p1.context-d1.icon.tau.ci95.1\endcsname{0}
\expandafter\def\csname pct@rev.desc.d7.sspro.tau-T-by-type.p1.context-d1.icon.tau.ci95.1\endcsname{45.0}
\expandafter\def\csname ppt@rev.desc.d7.sspro.tau-T-by-type.p1.context-d1.icon.tau.ci95.1\endcsname{$+$45.0}
\expandafter\def\csname val@rev.desc.d7.sspro.tau-T-by-type.p1.context-d1.icon.tau.mean\endcsname{0.200}
\expandafter\def\csname cnt@rev.desc.d7.sspro.tau-T-by-type.p1.context-d1.icon.tau.mean\endcsname{0}
\expandafter\def\csname pct@rev.desc.d7.sspro.tau-T-by-type.p1.context-d1.icon.tau.mean\endcsname{20.0}
\expandafter\def\csname ppt@rev.desc.d7.sspro.tau-T-by-type.p1.context-d1.icon.tau.mean\endcsname{$+$20.0}
\expandafter\def\csname val@rev.desc.d7.sspro.tau-T-by-type.p1.context-d1.icon.tau.n-apps\endcsname{10}
\expandafter\def\csname cnt@rev.desc.d7.sspro.tau-T-by-type.p1.context-d1.icon.tau.n-apps\endcsname{10}
\expandafter\def\csname pct@rev.desc.d7.sspro.tau-T-by-type.p1.context-d1.icon.tau.n-apps\endcsname{1000.0}
\expandafter\def\csname ppt@rev.desc.d7.sspro.tau-T-by-type.p1.context-d1.icon.tau.n-apps\endcsname{$+$1000.0}
\expandafter\def\csname val@rev.desc.d7.sspro.tau-T-by-type.p1.context-d1.icon.tau.tost1.ci90-inside\endcsname{0}
\expandafter\def\csname cnt@rev.desc.d7.sspro.tau-T-by-type.p1.context-d1.icon.tau.tost1.ci90-inside\endcsname{0}
\expandafter\def\csname pct@rev.desc.d7.sspro.tau-T-by-type.p1.context-d1.icon.tau.tost1.ci90-inside\endcsname{0.0}
\expandafter\def\csname ppt@rev.desc.d7.sspro.tau-T-by-type.p1.context-d1.icon.tau.tost1.ci90-inside\endcsname{0.0}
\expandafter\def\csname val@rev.desc.d7.sspro.tau-T-by-type.p1.context-d1.icon.tau.tost1.equivalent\endcsname{0}
\expandafter\def\csname cnt@rev.desc.d7.sspro.tau-T-by-type.p1.context-d1.icon.tau.tost1.equivalent\endcsname{0}
\expandafter\def\csname pct@rev.desc.d7.sspro.tau-T-by-type.p1.context-d1.icon.tau.tost1.equivalent\endcsname{0.0}
\expandafter\def\csname ppt@rev.desc.d7.sspro.tau-T-by-type.p1.context-d1.icon.tau.tost1.equivalent\endcsname{0.0}
\expandafter\def\csname val@rev.desc.d7.sspro.tau-T-by-type.p1.context-d1.icon.tau.tost1.p\endcsname{0.950}
\expandafter\def\csname cnt@rev.desc.d7.sspro.tau-T-by-type.p1.context-d1.icon.tau.tost1.p\endcsname{1}
\expandafter\def\csname pct@rev.desc.d7.sspro.tau-T-by-type.p1.context-d1.icon.tau.tost1.p\endcsname{95.0}
\expandafter\def\csname ppt@rev.desc.d7.sspro.tau-T-by-type.p1.context-d1.icon.tau.tost1.p\endcsname{$+$95.0}
\expandafter\def\csname val@rev.desc.d7.sspro.tau-T-by-type.p1.context-d1.icon.tau.tost5.ci90-inside\endcsname{0}
\expandafter\def\csname cnt@rev.desc.d7.sspro.tau-T-by-type.p1.context-d1.icon.tau.tost5.ci90-inside\endcsname{0}
\expandafter\def\csname pct@rev.desc.d7.sspro.tau-T-by-type.p1.context-d1.icon.tau.tost5.ci90-inside\endcsname{0.0}
\expandafter\def\csname ppt@rev.desc.d7.sspro.tau-T-by-type.p1.context-d1.icon.tau.tost5.ci90-inside\endcsname{0.0}
\expandafter\def\csname val@rev.desc.d7.sspro.tau-T-by-type.p1.context-d1.icon.tau.tost5.equivalent\endcsname{0}
\expandafter\def\csname cnt@rev.desc.d7.sspro.tau-T-by-type.p1.context-d1.icon.tau.tost5.equivalent\endcsname{0}
\expandafter\def\csname pct@rev.desc.d7.sspro.tau-T-by-type.p1.context-d1.icon.tau.tost5.equivalent\endcsname{0.0}
\expandafter\def\csname ppt@rev.desc.d7.sspro.tau-T-by-type.p1.context-d1.icon.tau.tost5.equivalent\endcsname{0.0}
\expandafter\def\csname val@rev.desc.d7.sspro.tau-T-by-type.p1.context-d1.icon.tau.tost5.p\endcsname{0.898}
\expandafter\def\csname cnt@rev.desc.d7.sspro.tau-T-by-type.p1.context-d1.icon.tau.tost5.p\endcsname{1}
\expandafter\def\csname pct@rev.desc.d7.sspro.tau-T-by-type.p1.context-d1.icon.tau.tost5.p\endcsname{89.8}
\expandafter\def\csname ppt@rev.desc.d7.sspro.tau-T-by-type.p1.context-d1.icon.tau.tost5.p\endcsname{$+$89.8}
\expandafter\def\csname val@rev.desc.d7.sspro.tau-T-by-type.p1.context-d1.text.acc-F\endcsname{0.897}
\expandafter\def\csname cnt@rev.desc.d7.sspro.tau-T-by-type.p1.context-d1.text.acc-F\endcsname{1}
\expandafter\def\csname pct@rev.desc.d7.sspro.tau-T-by-type.p1.context-d1.text.acc-F\endcsname{89.7}
\expandafter\def\csname ppt@rev.desc.d7.sspro.tau-T-by-type.p1.context-d1.text.acc-F\endcsname{$+$89.7}
\expandafter\def\csname val@rev.desc.d7.sspro.tau-T-by-type.p1.context-d1.text.acc-direct\endcsname{0.779}
\expandafter\def\csname cnt@rev.desc.d7.sspro.tau-T-by-type.p1.context-d1.text.acc-direct\endcsname{1}
\expandafter\def\csname pct@rev.desc.d7.sspro.tau-T-by-type.p1.context-d1.text.acc-direct\endcsname{77.9}
\expandafter\def\csname ppt@rev.desc.d7.sspro.tau-T-by-type.p1.context-d1.text.acc-direct\endcsname{$+$77.9}
\expandafter\def\csname val@rev.desc.d7.sspro.tau-T-by-type.p1.context-d1.text.n-apps\endcsname{12}
\expandafter\def\csname cnt@rev.desc.d7.sspro.tau-T-by-type.p1.context-d1.text.n-apps\endcsname{12}
\expandafter\def\csname pct@rev.desc.d7.sspro.tau-T-by-type.p1.context-d1.text.n-apps\endcsname{1200.0}
\expandafter\def\csname ppt@rev.desc.d7.sspro.tau-T-by-type.p1.context-d1.text.n-apps\endcsname{$+$1200.0}
\expandafter\def\csname val@rev.desc.d7.sspro.tau-T-by-type.p1.context-d1.text.n-tasks\endcsname{68}
\expandafter\def\csname cnt@rev.desc.d7.sspro.tau-T-by-type.p1.context-d1.text.n-tasks\endcsname{68}
\expandafter\def\csname pct@rev.desc.d7.sspro.tau-T-by-type.p1.context-d1.text.n-tasks\endcsname{6800.0}
\expandafter\def\csname ppt@rev.desc.d7.sspro.tau-T-by-type.p1.context-d1.text.n-tasks\endcsname{$+$6800.0}
\expandafter\def\csname val@rev.desc.d7.sspro.tau-T-by-type.p1.context-d1.text.tau.ci90.0\endcsname{-0.014}
\expandafter\def\csname cnt@rev.desc.d7.sspro.tau-T-by-type.p1.context-d1.text.tau.ci90.0\endcsname{0}
\expandafter\def\csname pct@rev.desc.d7.sspro.tau-T-by-type.p1.context-d1.text.tau.ci90.0\endcsname{$-$1.4}
\expandafter\def\csname ppt@rev.desc.d7.sspro.tau-T-by-type.p1.context-d1.text.tau.ci90.0\endcsname{$-$1.4}
\expandafter\def\csname val@rev.desc.d7.sspro.tau-T-by-type.p1.context-d1.text.tau.ci90.1\endcsname{0.202}
\expandafter\def\csname cnt@rev.desc.d7.sspro.tau-T-by-type.p1.context-d1.text.tau.ci90.1\endcsname{0}
\expandafter\def\csname pct@rev.desc.d7.sspro.tau-T-by-type.p1.context-d1.text.tau.ci90.1\endcsname{20.2}
\expandafter\def\csname ppt@rev.desc.d7.sspro.tau-T-by-type.p1.context-d1.text.tau.ci90.1\endcsname{$+$20.2}
\expandafter\def\csname val@rev.desc.d7.sspro.tau-T-by-type.p1.context-d1.text.tau.ci95-excludes-0\endcsname{0}
\expandafter\def\csname cnt@rev.desc.d7.sspro.tau-T-by-type.p1.context-d1.text.tau.ci95-excludes-0\endcsname{0}
\expandafter\def\csname pct@rev.desc.d7.sspro.tau-T-by-type.p1.context-d1.text.tau.ci95-excludes-0\endcsname{0.0}
\expandafter\def\csname ppt@rev.desc.d7.sspro.tau-T-by-type.p1.context-d1.text.tau.ci95-excludes-0\endcsname{0.0}
\expandafter\def\csname val@rev.desc.d7.sspro.tau-T-by-type.p1.context-d1.text.tau.ci95.0\endcsname{-0.027}
\expandafter\def\csname cnt@rev.desc.d7.sspro.tau-T-by-type.p1.context-d1.text.tau.ci95.0\endcsname{0}
\expandafter\def\csname pct@rev.desc.d7.sspro.tau-T-by-type.p1.context-d1.text.tau.ci95.0\endcsname{$-$2.7}
\expandafter\def\csname ppt@rev.desc.d7.sspro.tau-T-by-type.p1.context-d1.text.tau.ci95.0\endcsname{$-$2.7}
\expandafter\def\csname val@rev.desc.d7.sspro.tau-T-by-type.p1.context-d1.text.tau.ci95.1\endcsname{0.226}
\expandafter\def\csname cnt@rev.desc.d7.sspro.tau-T-by-type.p1.context-d1.text.tau.ci95.1\endcsname{0}
\expandafter\def\csname pct@rev.desc.d7.sspro.tau-T-by-type.p1.context-d1.text.tau.ci95.1\endcsname{22.6}
\expandafter\def\csname ppt@rev.desc.d7.sspro.tau-T-by-type.p1.context-d1.text.tau.ci95.1\endcsname{$+$22.6}
\expandafter\def\csname val@rev.desc.d7.sspro.tau-T-by-type.p1.context-d1.text.tau.mean\endcsname{0.086}
\expandafter\def\csname cnt@rev.desc.d7.sspro.tau-T-by-type.p1.context-d1.text.tau.mean\endcsname{0}
\expandafter\def\csname pct@rev.desc.d7.sspro.tau-T-by-type.p1.context-d1.text.tau.mean\endcsname{8.6}
\expandafter\def\csname ppt@rev.desc.d7.sspro.tau-T-by-type.p1.context-d1.text.tau.mean\endcsname{$+$8.6}
\expandafter\def\csname val@rev.desc.d7.sspro.tau-T-by-type.p1.context-d1.text.tau.n-apps\endcsname{12}
\expandafter\def\csname cnt@rev.desc.d7.sspro.tau-T-by-type.p1.context-d1.text.tau.n-apps\endcsname{12}
\expandafter\def\csname pct@rev.desc.d7.sspro.tau-T-by-type.p1.context-d1.text.tau.n-apps\endcsname{1200.0}
\expandafter\def\csname ppt@rev.desc.d7.sspro.tau-T-by-type.p1.context-d1.text.tau.n-apps\endcsname{$+$1200.0}
\expandafter\def\csname val@rev.desc.d7.sspro.tau-T-by-type.p1.context-d1.text.tau.tost1.ci90-inside\endcsname{0}
\expandafter\def\csname cnt@rev.desc.d7.sspro.tau-T-by-type.p1.context-d1.text.tau.tost1.ci90-inside\endcsname{0}
\expandafter\def\csname pct@rev.desc.d7.sspro.tau-T-by-type.p1.context-d1.text.tau.tost1.ci90-inside\endcsname{0.0}
\expandafter\def\csname ppt@rev.desc.d7.sspro.tau-T-by-type.p1.context-d1.text.tau.tost1.ci90-inside\endcsname{0.0}
\expandafter\def\csname val@rev.desc.d7.sspro.tau-T-by-type.p1.context-d1.text.tau.tost1.equivalent\endcsname{0}
\expandafter\def\csname cnt@rev.desc.d7.sspro.tau-T-by-type.p1.context-d1.text.tau.tost1.equivalent\endcsname{0}
\expandafter\def\csname pct@rev.desc.d7.sspro.tau-T-by-type.p1.context-d1.text.tau.tost1.equivalent\endcsname{0.0}
\expandafter\def\csname ppt@rev.desc.d7.sspro.tau-T-by-type.p1.context-d1.text.tau.tost1.equivalent\endcsname{0.0}
\expandafter\def\csname val@rev.desc.d7.sspro.tau-T-by-type.p1.context-d1.text.tau.tost1.p\endcsname{0.882}
\expandafter\def\csname cnt@rev.desc.d7.sspro.tau-T-by-type.p1.context-d1.text.tau.tost1.p\endcsname{1}
\expandafter\def\csname pct@rev.desc.d7.sspro.tau-T-by-type.p1.context-d1.text.tau.tost1.p\endcsname{88.2}
\expandafter\def\csname ppt@rev.desc.d7.sspro.tau-T-by-type.p1.context-d1.text.tau.tost1.p\endcsname{$+$88.2}
\expandafter\def\csname val@rev.desc.d7.sspro.tau-T-by-type.p1.context-d1.text.tau.tost5.ci90-inside\endcsname{0}
\expandafter\def\csname cnt@rev.desc.d7.sspro.tau-T-by-type.p1.context-d1.text.tau.tost5.ci90-inside\endcsname{0}
\expandafter\def\csname pct@rev.desc.d7.sspro.tau-T-by-type.p1.context-d1.text.tau.tost5.ci90-inside\endcsname{0.0}
\expandafter\def\csname ppt@rev.desc.d7.sspro.tau-T-by-type.p1.context-d1.text.tau.tost5.ci90-inside\endcsname{0.0}
\expandafter\def\csname val@rev.desc.d7.sspro.tau-T-by-type.p1.context-d1.text.tau.tost5.equivalent\endcsname{0}
\expandafter\def\csname cnt@rev.desc.d7.sspro.tau-T-by-type.p1.context-d1.text.tau.tost5.equivalent\endcsname{0}
\expandafter\def\csname pct@rev.desc.d7.sspro.tau-T-by-type.p1.context-d1.text.tau.tost5.equivalent\endcsname{0.0}
\expandafter\def\csname ppt@rev.desc.d7.sspro.tau-T-by-type.p1.context-d1.text.tau.tost5.equivalent\endcsname{0.0}
\expandafter\def\csname val@rev.desc.d7.sspro.tau-T-by-type.p1.context-d1.text.tau.tost5.p\endcsname{0.672}
\expandafter\def\csname cnt@rev.desc.d7.sspro.tau-T-by-type.p1.context-d1.text.tau.tost5.p\endcsname{1}
\expandafter\def\csname pct@rev.desc.d7.sspro.tau-T-by-type.p1.context-d1.text.tau.tost5.p\endcsname{67.2}
\expandafter\def\csname ppt@rev.desc.d7.sspro.tau-T-by-type.p1.context-d1.text.tau.tost5.p\endcsname{$+$67.2}
\expandafter\def\csname val@rev.desc.d7.sspro.tau-T-by-type.p1.dual-d2.icon-minus-text-on-apps-with-both.ci90.0\endcsname{0.030}
\expandafter\def\csname cnt@rev.desc.d7.sspro.tau-T-by-type.p1.dual-d2.icon-minus-text-on-apps-with-both.ci90.0\endcsname{0}
\expandafter\def\csname pct@rev.desc.d7.sspro.tau-T-by-type.p1.dual-d2.icon-minus-text-on-apps-with-both.ci90.0\endcsname{3.0}
\expandafter\def\csname ppt@rev.desc.d7.sspro.tau-T-by-type.p1.dual-d2.icon-minus-text-on-apps-with-both.ci90.0\endcsname{$+$3.0}
\expandafter\def\csname val@rev.desc.d7.sspro.tau-T-by-type.p1.dual-d2.icon-minus-text-on-apps-with-both.ci90.1\endcsname{0.453}
\expandafter\def\csname cnt@rev.desc.d7.sspro.tau-T-by-type.p1.dual-d2.icon-minus-text-on-apps-with-both.ci90.1\endcsname{0}
\expandafter\def\csname pct@rev.desc.d7.sspro.tau-T-by-type.p1.dual-d2.icon-minus-text-on-apps-with-both.ci90.1\endcsname{45.3}
\expandafter\def\csname ppt@rev.desc.d7.sspro.tau-T-by-type.p1.dual-d2.icon-minus-text-on-apps-with-both.ci90.1\endcsname{$+$45.3}
\expandafter\def\csname val@rev.desc.d7.sspro.tau-T-by-type.p1.dual-d2.icon-minus-text-on-apps-with-both.ci95-excludes-0\endcsname{0}
\expandafter\def\csname cnt@rev.desc.d7.sspro.tau-T-by-type.p1.dual-d2.icon-minus-text-on-apps-with-both.ci95-excludes-0\endcsname{0}
\expandafter\def\csname pct@rev.desc.d7.sspro.tau-T-by-type.p1.dual-d2.icon-minus-text-on-apps-with-both.ci95-excludes-0\endcsname{0.0}
\expandafter\def\csname ppt@rev.desc.d7.sspro.tau-T-by-type.p1.dual-d2.icon-minus-text-on-apps-with-both.ci95-excludes-0\endcsname{0.0}
\expandafter\def\csname val@rev.desc.d7.sspro.tau-T-by-type.p1.dual-d2.icon-minus-text-on-apps-with-both.ci95.0\endcsname{-0.002}
\expandafter\def\csname cnt@rev.desc.d7.sspro.tau-T-by-type.p1.dual-d2.icon-minus-text-on-apps-with-both.ci95.0\endcsname{0}
\expandafter\def\csname pct@rev.desc.d7.sspro.tau-T-by-type.p1.dual-d2.icon-minus-text-on-apps-with-both.ci95.0\endcsname{$-$0.2}
\expandafter\def\csname ppt@rev.desc.d7.sspro.tau-T-by-type.p1.dual-d2.icon-minus-text-on-apps-with-both.ci95.0\endcsname{$-$0.2}
\expandafter\def\csname val@rev.desc.d7.sspro.tau-T-by-type.p1.dual-d2.icon-minus-text-on-apps-with-both.ci95.1\endcsname{0.510}
\expandafter\def\csname cnt@rev.desc.d7.sspro.tau-T-by-type.p1.dual-d2.icon-minus-text-on-apps-with-both.ci95.1\endcsname{1}
\expandafter\def\csname pct@rev.desc.d7.sspro.tau-T-by-type.p1.dual-d2.icon-minus-text-on-apps-with-both.ci95.1\endcsname{51.0}
\expandafter\def\csname ppt@rev.desc.d7.sspro.tau-T-by-type.p1.dual-d2.icon-minus-text-on-apps-with-both.ci95.1\endcsname{$+$51.0}
\expandafter\def\csname val@rev.desc.d7.sspro.tau-T-by-type.p1.dual-d2.icon-minus-text-on-apps-with-both.mean\endcsname{0.230}
\expandafter\def\csname cnt@rev.desc.d7.sspro.tau-T-by-type.p1.dual-d2.icon-minus-text-on-apps-with-both.mean\endcsname{0}
\expandafter\def\csname pct@rev.desc.d7.sspro.tau-T-by-type.p1.dual-d2.icon-minus-text-on-apps-with-both.mean\endcsname{23.0}
\expandafter\def\csname ppt@rev.desc.d7.sspro.tau-T-by-type.p1.dual-d2.icon-minus-text-on-apps-with-both.mean\endcsname{$+$23.0}
\expandafter\def\csname val@rev.desc.d7.sspro.tau-T-by-type.p1.dual-d2.icon-minus-text-on-apps-with-both.n-apps\endcsname{10}
\expandafter\def\csname cnt@rev.desc.d7.sspro.tau-T-by-type.p1.dual-d2.icon-minus-text-on-apps-with-both.n-apps\endcsname{10}
\expandafter\def\csname pct@rev.desc.d7.sspro.tau-T-by-type.p1.dual-d2.icon-minus-text-on-apps-with-both.n-apps\endcsname{1000.0}
\expandafter\def\csname ppt@rev.desc.d7.sspro.tau-T-by-type.p1.dual-d2.icon-minus-text-on-apps-with-both.n-apps\endcsname{$+$1000.0}
\expandafter\def\csname val@rev.desc.d7.sspro.tau-T-by-type.p1.dual-d2.icon-minus-text-on-apps-with-both.tost1.ci90-inside\endcsname{0}
\expandafter\def\csname cnt@rev.desc.d7.sspro.tau-T-by-type.p1.dual-d2.icon-minus-text-on-apps-with-both.tost1.ci90-inside\endcsname{0}
\expandafter\def\csname pct@rev.desc.d7.sspro.tau-T-by-type.p1.dual-d2.icon-minus-text-on-apps-with-both.tost1.ci90-inside\endcsname{0.0}
\expandafter\def\csname ppt@rev.desc.d7.sspro.tau-T-by-type.p1.dual-d2.icon-minus-text-on-apps-with-both.tost1.ci90-inside\endcsname{0.0}
\expandafter\def\csname val@rev.desc.d7.sspro.tau-T-by-type.p1.dual-d2.icon-minus-text-on-apps-with-both.tost1.equivalent\endcsname{0}
\expandafter\def\csname cnt@rev.desc.d7.sspro.tau-T-by-type.p1.dual-d2.icon-minus-text-on-apps-with-both.tost1.equivalent\endcsname{0}
\expandafter\def\csname pct@rev.desc.d7.sspro.tau-T-by-type.p1.dual-d2.icon-minus-text-on-apps-with-both.tost1.equivalent\endcsname{0.0}
\expandafter\def\csname ppt@rev.desc.d7.sspro.tau-T-by-type.p1.dual-d2.icon-minus-text-on-apps-with-both.tost1.equivalent\endcsname{0.0}
\expandafter\def\csname val@rev.desc.d7.sspro.tau-T-by-type.p1.dual-d2.icon-minus-text-on-apps-with-both.tost1.p\endcsname{0.964}
\expandafter\def\csname cnt@rev.desc.d7.sspro.tau-T-by-type.p1.dual-d2.icon-minus-text-on-apps-with-both.tost1.p\endcsname{1}
\expandafter\def\csname pct@rev.desc.d7.sspro.tau-T-by-type.p1.dual-d2.icon-minus-text-on-apps-with-both.tost1.p\endcsname{96.4}
\expandafter\def\csname ppt@rev.desc.d7.sspro.tau-T-by-type.p1.dual-d2.icon-minus-text-on-apps-with-both.tost1.p\endcsname{$+$96.4}
\expandafter\def\csname val@rev.desc.d7.sspro.tau-T-by-type.p1.dual-d2.icon-minus-text-on-apps-with-both.tost5.ci90-inside\endcsname{0}
\expandafter\def\csname cnt@rev.desc.d7.sspro.tau-T-by-type.p1.dual-d2.icon-minus-text-on-apps-with-both.tost5.ci90-inside\endcsname{0}
\expandafter\def\csname pct@rev.desc.d7.sspro.tau-T-by-type.p1.dual-d2.icon-minus-text-on-apps-with-both.tost5.ci90-inside\endcsname{0.0}
\expandafter\def\csname ppt@rev.desc.d7.sspro.tau-T-by-type.p1.dual-d2.icon-minus-text-on-apps-with-both.tost5.ci90-inside\endcsname{0.0}
\expandafter\def\csname val@rev.desc.d7.sspro.tau-T-by-type.p1.dual-d2.icon-minus-text-on-apps-with-both.tost5.equivalent\endcsname{0}
\expandafter\def\csname cnt@rev.desc.d7.sspro.tau-T-by-type.p1.dual-d2.icon-minus-text-on-apps-with-both.tost5.equivalent\endcsname{0}
\expandafter\def\csname pct@rev.desc.d7.sspro.tau-T-by-type.p1.dual-d2.icon-minus-text-on-apps-with-both.tost5.equivalent\endcsname{0.0}
\expandafter\def\csname ppt@rev.desc.d7.sspro.tau-T-by-type.p1.dual-d2.icon-minus-text-on-apps-with-both.tost5.equivalent\endcsname{0.0}
\expandafter\def\csname val@rev.desc.d7.sspro.tau-T-by-type.p1.dual-d2.icon-minus-text-on-apps-with-both.tost5.p\endcsname{0.926}
\expandafter\def\csname cnt@rev.desc.d7.sspro.tau-T-by-type.p1.dual-d2.icon-minus-text-on-apps-with-both.tost5.p\endcsname{1}
\expandafter\def\csname pct@rev.desc.d7.sspro.tau-T-by-type.p1.dual-d2.icon-minus-text-on-apps-with-both.tost5.p\endcsname{92.6}
\expandafter\def\csname ppt@rev.desc.d7.sspro.tau-T-by-type.p1.dual-d2.icon-minus-text-on-apps-with-both.tost5.p\endcsname{$+$92.6}
\expandafter\def\csname val@rev.desc.d7.sspro.tau-T-by-type.p1.dual-d2.icon.acc-F\endcsname{0.786}
\expandafter\def\csname cnt@rev.desc.d7.sspro.tau-T-by-type.p1.dual-d2.icon.acc-F\endcsname{1}
\expandafter\def\csname pct@rev.desc.d7.sspro.tau-T-by-type.p1.dual-d2.icon.acc-F\endcsname{78.6}
\expandafter\def\csname ppt@rev.desc.d7.sspro.tau-T-by-type.p1.dual-d2.icon.acc-F\endcsname{$+$78.6}
\expandafter\def\csname val@rev.desc.d7.sspro.tau-T-by-type.p1.dual-d2.icon.acc-direct\endcsname{0.643}
\expandafter\def\csname cnt@rev.desc.d7.sspro.tau-T-by-type.p1.dual-d2.icon.acc-direct\endcsname{1}
\expandafter\def\csname pct@rev.desc.d7.sspro.tau-T-by-type.p1.dual-d2.icon.acc-direct\endcsname{64.3}
\expandafter\def\csname ppt@rev.desc.d7.sspro.tau-T-by-type.p1.dual-d2.icon.acc-direct\endcsname{$+$64.3}
\expandafter\def\csname val@rev.desc.d7.sspro.tau-T-by-type.p1.dual-d2.icon.n-apps\endcsname{10}
\expandafter\def\csname cnt@rev.desc.d7.sspro.tau-T-by-type.p1.dual-d2.icon.n-apps\endcsname{10}
\expandafter\def\csname pct@rev.desc.d7.sspro.tau-T-by-type.p1.dual-d2.icon.n-apps\endcsname{1000.0}
\expandafter\def\csname ppt@rev.desc.d7.sspro.tau-T-by-type.p1.dual-d2.icon.n-apps\endcsname{$+$1000.0}
\expandafter\def\csname val@rev.desc.d7.sspro.tau-T-by-type.p1.dual-d2.icon.n-tasks\endcsname{28}
\expandafter\def\csname cnt@rev.desc.d7.sspro.tau-T-by-type.p1.dual-d2.icon.n-tasks\endcsname{28}
\expandafter\def\csname pct@rev.desc.d7.sspro.tau-T-by-type.p1.dual-d2.icon.n-tasks\endcsname{2800.0}
\expandafter\def\csname ppt@rev.desc.d7.sspro.tau-T-by-type.p1.dual-d2.icon.n-tasks\endcsname{$+$2800.0}
\expandafter\def\csname val@rev.desc.d7.sspro.tau-T-by-type.p1.dual-d2.icon.tau.ci90.0\endcsname{0.033}
\expandafter\def\csname cnt@rev.desc.d7.sspro.tau-T-by-type.p1.dual-d2.icon.tau.ci90.0\endcsname{0}
\expandafter\def\csname pct@rev.desc.d7.sspro.tau-T-by-type.p1.dual-d2.icon.tau.ci90.0\endcsname{3.3}
\expandafter\def\csname ppt@rev.desc.d7.sspro.tau-T-by-type.p1.dual-d2.icon.tau.ci90.0\endcsname{$+$3.3}
\expandafter\def\csname val@rev.desc.d7.sspro.tau-T-by-type.p1.dual-d2.icon.tau.ci90.1\endcsname{0.447}
\expandafter\def\csname cnt@rev.desc.d7.sspro.tau-T-by-type.p1.dual-d2.icon.tau.ci90.1\endcsname{0}
\expandafter\def\csname pct@rev.desc.d7.sspro.tau-T-by-type.p1.dual-d2.icon.tau.ci90.1\endcsname{44.7}
\expandafter\def\csname ppt@rev.desc.d7.sspro.tau-T-by-type.p1.dual-d2.icon.tau.ci90.1\endcsname{$+$44.7}
\expandafter\def\csname val@rev.desc.d7.sspro.tau-T-by-type.p1.dual-d2.icon.tau.ci95-excludes-0\endcsname{1}
\expandafter\def\csname cnt@rev.desc.d7.sspro.tau-T-by-type.p1.dual-d2.icon.tau.ci95-excludes-0\endcsname{1}
\expandafter\def\csname pct@rev.desc.d7.sspro.tau-T-by-type.p1.dual-d2.icon.tau.ci95-excludes-0\endcsname{100.0}
\expandafter\def\csname ppt@rev.desc.d7.sspro.tau-T-by-type.p1.dual-d2.icon.tau.ci95-excludes-0\endcsname{$+$100.0}
\expandafter\def\csname val@rev.desc.d7.sspro.tau-T-by-type.p1.dual-d2.icon.tau.ci95.0\endcsname{0.007}
\expandafter\def\csname cnt@rev.desc.d7.sspro.tau-T-by-type.p1.dual-d2.icon.tau.ci95.0\endcsname{0}
\expandafter\def\csname pct@rev.desc.d7.sspro.tau-T-by-type.p1.dual-d2.icon.tau.ci95.0\endcsname{0.7}
\expandafter\def\csname ppt@rev.desc.d7.sspro.tau-T-by-type.p1.dual-d2.icon.tau.ci95.0\endcsname{$+$0.7}
\expandafter\def\csname val@rev.desc.d7.sspro.tau-T-by-type.p1.dual-d2.icon.tau.ci95.1\endcsname{0.487}
\expandafter\def\csname cnt@rev.desc.d7.sspro.tau-T-by-type.p1.dual-d2.icon.tau.ci95.1\endcsname{0}
\expandafter\def\csname pct@rev.desc.d7.sspro.tau-T-by-type.p1.dual-d2.icon.tau.ci95.1\endcsname{48.7}
\expandafter\def\csname ppt@rev.desc.d7.sspro.tau-T-by-type.p1.dual-d2.icon.tau.ci95.1\endcsname{$+$48.7}
\expandafter\def\csname val@rev.desc.d7.sspro.tau-T-by-type.p1.dual-d2.icon.tau.mean\endcsname{0.233}
\expandafter\def\csname cnt@rev.desc.d7.sspro.tau-T-by-type.p1.dual-d2.icon.tau.mean\endcsname{0}
\expandafter\def\csname pct@rev.desc.d7.sspro.tau-T-by-type.p1.dual-d2.icon.tau.mean\endcsname{23.3}
\expandafter\def\csname ppt@rev.desc.d7.sspro.tau-T-by-type.p1.dual-d2.icon.tau.mean\endcsname{$+$23.3}
\expandafter\def\csname val@rev.desc.d7.sspro.tau-T-by-type.p1.dual-d2.icon.tau.n-apps\endcsname{10}
\expandafter\def\csname cnt@rev.desc.d7.sspro.tau-T-by-type.p1.dual-d2.icon.tau.n-apps\endcsname{10}
\expandafter\def\csname pct@rev.desc.d7.sspro.tau-T-by-type.p1.dual-d2.icon.tau.n-apps\endcsname{1000.0}
\expandafter\def\csname ppt@rev.desc.d7.sspro.tau-T-by-type.p1.dual-d2.icon.tau.n-apps\endcsname{$+$1000.0}
\expandafter\def\csname val@rev.desc.d7.sspro.tau-T-by-type.p1.dual-d2.icon.tau.tost1.ci90-inside\endcsname{0}
\expandafter\def\csname cnt@rev.desc.d7.sspro.tau-T-by-type.p1.dual-d2.icon.tau.tost1.ci90-inside\endcsname{0}
\expandafter\def\csname pct@rev.desc.d7.sspro.tau-T-by-type.p1.dual-d2.icon.tau.tost1.ci90-inside\endcsname{0.0}
\expandafter\def\csname ppt@rev.desc.d7.sspro.tau-T-by-type.p1.dual-d2.icon.tau.tost1.ci90-inside\endcsname{0.0}
\expandafter\def\csname val@rev.desc.d7.sspro.tau-T-by-type.p1.dual-d2.icon.tau.tost1.equivalent\endcsname{0}
\expandafter\def\csname cnt@rev.desc.d7.sspro.tau-T-by-type.p1.dual-d2.icon.tau.tost1.equivalent\endcsname{0}
\expandafter\def\csname pct@rev.desc.d7.sspro.tau-T-by-type.p1.dual-d2.icon.tau.tost1.equivalent\endcsname{0.0}
\expandafter\def\csname ppt@rev.desc.d7.sspro.tau-T-by-type.p1.dual-d2.icon.tau.tost1.equivalent\endcsname{0.0}
\expandafter\def\csname val@rev.desc.d7.sspro.tau-T-by-type.p1.dual-d2.icon.tau.tost1.p\endcsname{0.974}
\expandafter\def\csname cnt@rev.desc.d7.sspro.tau-T-by-type.p1.dual-d2.icon.tau.tost1.p\endcsname{1}
\expandafter\def\csname pct@rev.desc.d7.sspro.tau-T-by-type.p1.dual-d2.icon.tau.tost1.p\endcsname{97.4}
\expandafter\def\csname ppt@rev.desc.d7.sspro.tau-T-by-type.p1.dual-d2.icon.tau.tost1.p\endcsname{$+$97.4}
\expandafter\def\csname val@rev.desc.d7.sspro.tau-T-by-type.p1.dual-d2.icon.tau.tost5.ci90-inside\endcsname{0}
\expandafter\def\csname cnt@rev.desc.d7.sspro.tau-T-by-type.p1.dual-d2.icon.tau.tost5.ci90-inside\endcsname{0}
\expandafter\def\csname pct@rev.desc.d7.sspro.tau-T-by-type.p1.dual-d2.icon.tau.tost5.ci90-inside\endcsname{0.0}
\expandafter\def\csname ppt@rev.desc.d7.sspro.tau-T-by-type.p1.dual-d2.icon.tau.tost5.ci90-inside\endcsname{0.0}
\expandafter\def\csname val@rev.desc.d7.sspro.tau-T-by-type.p1.dual-d2.icon.tau.tost5.equivalent\endcsname{0}
\expandafter\def\csname cnt@rev.desc.d7.sspro.tau-T-by-type.p1.dual-d2.icon.tau.tost5.equivalent\endcsname{0}
\expandafter\def\csname pct@rev.desc.d7.sspro.tau-T-by-type.p1.dual-d2.icon.tau.tost5.equivalent\endcsname{0.0}
\expandafter\def\csname ppt@rev.desc.d7.sspro.tau-T-by-type.p1.dual-d2.icon.tau.tost5.equivalent\endcsname{0.0}
\expandafter\def\csname val@rev.desc.d7.sspro.tau-T-by-type.p1.dual-d2.icon.tau.tost5.p\endcsname{0.933}
\expandafter\def\csname cnt@rev.desc.d7.sspro.tau-T-by-type.p1.dual-d2.icon.tau.tost5.p\endcsname{1}
\expandafter\def\csname pct@rev.desc.d7.sspro.tau-T-by-type.p1.dual-d2.icon.tau.tost5.p\endcsname{93.3}
\expandafter\def\csname ppt@rev.desc.d7.sspro.tau-T-by-type.p1.dual-d2.icon.tau.tost5.p\endcsname{$+$93.3}
\expandafter\def\csname val@rev.desc.d7.sspro.tau-T-by-type.p1.dual-d2.text.acc-F\endcsname{0.926}
\expandafter\def\csname cnt@rev.desc.d7.sspro.tau-T-by-type.p1.dual-d2.text.acc-F\endcsname{1}
\expandafter\def\csname pct@rev.desc.d7.sspro.tau-T-by-type.p1.dual-d2.text.acc-F\endcsname{92.6}
\expandafter\def\csname ppt@rev.desc.d7.sspro.tau-T-by-type.p1.dual-d2.text.acc-F\endcsname{$+$92.6}
\expandafter\def\csname val@rev.desc.d7.sspro.tau-T-by-type.p1.dual-d2.text.acc-direct\endcsname{0.779}
\expandafter\def\csname cnt@rev.desc.d7.sspro.tau-T-by-type.p1.dual-d2.text.acc-direct\endcsname{1}
\expandafter\def\csname pct@rev.desc.d7.sspro.tau-T-by-type.p1.dual-d2.text.acc-direct\endcsname{77.9}
\expandafter\def\csname ppt@rev.desc.d7.sspro.tau-T-by-type.p1.dual-d2.text.acc-direct\endcsname{$+$77.9}
\expandafter\def\csname val@rev.desc.d7.sspro.tau-T-by-type.p1.dual-d2.text.n-apps\endcsname{12}
\expandafter\def\csname cnt@rev.desc.d7.sspro.tau-T-by-type.p1.dual-d2.text.n-apps\endcsname{12}
\expandafter\def\csname pct@rev.desc.d7.sspro.tau-T-by-type.p1.dual-d2.text.n-apps\endcsname{1200.0}
\expandafter\def\csname ppt@rev.desc.d7.sspro.tau-T-by-type.p1.dual-d2.text.n-apps\endcsname{$+$1200.0}
\expandafter\def\csname val@rev.desc.d7.sspro.tau-T-by-type.p1.dual-d2.text.n-tasks\endcsname{68}
\expandafter\def\csname cnt@rev.desc.d7.sspro.tau-T-by-type.p1.dual-d2.text.n-tasks\endcsname{68}
\expandafter\def\csname pct@rev.desc.d7.sspro.tau-T-by-type.p1.dual-d2.text.n-tasks\endcsname{6800.0}
\expandafter\def\csname ppt@rev.desc.d7.sspro.tau-T-by-type.p1.dual-d2.text.n-tasks\endcsname{$+$6800.0}
\expandafter\def\csname val@rev.desc.d7.sspro.tau-T-by-type.p1.dual-d2.text.tau.ci90.0\endcsname{-0.010}
\expandafter\def\csname cnt@rev.desc.d7.sspro.tau-T-by-type.p1.dual-d2.text.tau.ci90.0\endcsname{0}
\expandafter\def\csname pct@rev.desc.d7.sspro.tau-T-by-type.p1.dual-d2.text.tau.ci90.0\endcsname{$-$1.0}
\expandafter\def\csname ppt@rev.desc.d7.sspro.tau-T-by-type.p1.dual-d2.text.tau.ci90.0\endcsname{$-$1.0}
\expandafter\def\csname val@rev.desc.d7.sspro.tau-T-by-type.p1.dual-d2.text.tau.ci90.1\endcsname{0.242}
\expandafter\def\csname cnt@rev.desc.d7.sspro.tau-T-by-type.p1.dual-d2.text.tau.ci90.1\endcsname{0}
\expandafter\def\csname pct@rev.desc.d7.sspro.tau-T-by-type.p1.dual-d2.text.tau.ci90.1\endcsname{24.2}
\expandafter\def\csname ppt@rev.desc.d7.sspro.tau-T-by-type.p1.dual-d2.text.tau.ci90.1\endcsname{$+$24.2}
\expandafter\def\csname val@rev.desc.d7.sspro.tau-T-by-type.p1.dual-d2.text.tau.ci95-excludes-0\endcsname{0}
\expandafter\def\csname cnt@rev.desc.d7.sspro.tau-T-by-type.p1.dual-d2.text.tau.ci95-excludes-0\endcsname{0}
\expandafter\def\csname pct@rev.desc.d7.sspro.tau-T-by-type.p1.dual-d2.text.tau.ci95-excludes-0\endcsname{0.0}
\expandafter\def\csname ppt@rev.desc.d7.sspro.tau-T-by-type.p1.dual-d2.text.tau.ci95-excludes-0\endcsname{0.0}
\expandafter\def\csname val@rev.desc.d7.sspro.tau-T-by-type.p1.dual-d2.text.tau.ci95.0\endcsname{-0.028}
\expandafter\def\csname cnt@rev.desc.d7.sspro.tau-T-by-type.p1.dual-d2.text.tau.ci95.0\endcsname{0}
\expandafter\def\csname pct@rev.desc.d7.sspro.tau-T-by-type.p1.dual-d2.text.tau.ci95.0\endcsname{$-$2.8}
\expandafter\def\csname ppt@rev.desc.d7.sspro.tau-T-by-type.p1.dual-d2.text.tau.ci95.0\endcsname{$-$2.8}
\expandafter\def\csname val@rev.desc.d7.sspro.tau-T-by-type.p1.dual-d2.text.tau.ci95.1\endcsname{0.270}
\expandafter\def\csname cnt@rev.desc.d7.sspro.tau-T-by-type.p1.dual-d2.text.tau.ci95.1\endcsname{0}
\expandafter\def\csname pct@rev.desc.d7.sspro.tau-T-by-type.p1.dual-d2.text.tau.ci95.1\endcsname{27.0}
\expandafter\def\csname ppt@rev.desc.d7.sspro.tau-T-by-type.p1.dual-d2.text.tau.ci95.1\endcsname{$+$27.0}
\expandafter\def\csname val@rev.desc.d7.sspro.tau-T-by-type.p1.dual-d2.text.tau.mean\endcsname{0.107}
\expandafter\def\csname cnt@rev.desc.d7.sspro.tau-T-by-type.p1.dual-d2.text.tau.mean\endcsname{0}
\expandafter\def\csname pct@rev.desc.d7.sspro.tau-T-by-type.p1.dual-d2.text.tau.mean\endcsname{10.7}
\expandafter\def\csname ppt@rev.desc.d7.sspro.tau-T-by-type.p1.dual-d2.text.tau.mean\endcsname{$+$10.7}
\expandafter\def\csname val@rev.desc.d7.sspro.tau-T-by-type.p1.dual-d2.text.tau.n-apps\endcsname{12}
\expandafter\def\csname cnt@rev.desc.d7.sspro.tau-T-by-type.p1.dual-d2.text.tau.n-apps\endcsname{12}
\expandafter\def\csname pct@rev.desc.d7.sspro.tau-T-by-type.p1.dual-d2.text.tau.n-apps\endcsname{1200.0}
\expandafter\def\csname ppt@rev.desc.d7.sspro.tau-T-by-type.p1.dual-d2.text.tau.n-apps\endcsname{$+$1200.0}
\expandafter\def\csname val@rev.desc.d7.sspro.tau-T-by-type.p1.dual-d2.text.tau.tost1.ci90-inside\endcsname{0}
\expandafter\def\csname cnt@rev.desc.d7.sspro.tau-T-by-type.p1.dual-d2.text.tau.tost1.ci90-inside\endcsname{0}
\expandafter\def\csname pct@rev.desc.d7.sspro.tau-T-by-type.p1.dual-d2.text.tau.tost1.ci90-inside\endcsname{0.0}
\expandafter\def\csname ppt@rev.desc.d7.sspro.tau-T-by-type.p1.dual-d2.text.tau.tost1.ci90-inside\endcsname{0.0}
\expandafter\def\csname val@rev.desc.d7.sspro.tau-T-by-type.p1.dual-d2.text.tau.tost1.equivalent\endcsname{0}
\expandafter\def\csname cnt@rev.desc.d7.sspro.tau-T-by-type.p1.dual-d2.text.tau.tost1.equivalent\endcsname{0}
\expandafter\def\csname pct@rev.desc.d7.sspro.tau-T-by-type.p1.dual-d2.text.tau.tost1.equivalent\endcsname{0.0}
\expandafter\def\csname ppt@rev.desc.d7.sspro.tau-T-by-type.p1.dual-d2.text.tau.tost1.equivalent\endcsname{0.0}
\expandafter\def\csname val@rev.desc.d7.sspro.tau-T-by-type.p1.dual-d2.text.tau.tost1.p\endcsname{0.902}
\expandafter\def\csname cnt@rev.desc.d7.sspro.tau-T-by-type.p1.dual-d2.text.tau.tost1.p\endcsname{1}
\expandafter\def\csname pct@rev.desc.d7.sspro.tau-T-by-type.p1.dual-d2.text.tau.tost1.p\endcsname{90.2}
\expandafter\def\csname ppt@rev.desc.d7.sspro.tau-T-by-type.p1.dual-d2.text.tau.tost1.p\endcsname{$+$90.2}
\expandafter\def\csname val@rev.desc.d7.sspro.tau-T-by-type.p1.dual-d2.text.tau.tost5.ci90-inside\endcsname{0}
\expandafter\def\csname cnt@rev.desc.d7.sspro.tau-T-by-type.p1.dual-d2.text.tau.tost5.ci90-inside\endcsname{0}
\expandafter\def\csname pct@rev.desc.d7.sspro.tau-T-by-type.p1.dual-d2.text.tau.tost5.ci90-inside\endcsname{0.0}
\expandafter\def\csname ppt@rev.desc.d7.sspro.tau-T-by-type.p1.dual-d2.text.tau.tost5.ci90-inside\endcsname{0.0}
\expandafter\def\csname val@rev.desc.d7.sspro.tau-T-by-type.p1.dual-d2.text.tau.tost5.equivalent\endcsname{0}
\expandafter\def\csname cnt@rev.desc.d7.sspro.tau-T-by-type.p1.dual-d2.text.tau.tost5.equivalent\endcsname{0}
\expandafter\def\csname pct@rev.desc.d7.sspro.tau-T-by-type.p1.dual-d2.text.tau.tost5.equivalent\endcsname{0.0}
\expandafter\def\csname ppt@rev.desc.d7.sspro.tau-T-by-type.p1.dual-d2.text.tau.tost5.equivalent\endcsname{0.0}
\expandafter\def\csname val@rev.desc.d7.sspro.tau-T-by-type.p1.dual-d2.text.tau.tost5.p\endcsname{0.764}
\expandafter\def\csname cnt@rev.desc.d7.sspro.tau-T-by-type.p1.dual-d2.text.tau.tost5.p\endcsname{1}
\expandafter\def\csname pct@rev.desc.d7.sspro.tau-T-by-type.p1.dual-d2.text.tau.tost5.p\endcsname{76.4}
\expandafter\def\csname ppt@rev.desc.d7.sspro.tau-T-by-type.p1.dual-d2.text.tau.tost5.p\endcsname{$+$76.4}
\expandafter\def\csname val@rev.desc.d7.sspro.tau-T-by-type.p1.grid2-d3.icon-minus-text-on-apps-with-both.ci90.0\endcsname{0.058}
\expandafter\def\csname cnt@rev.desc.d7.sspro.tau-T-by-type.p1.grid2-d3.icon-minus-text-on-apps-with-both.ci90.0\endcsname{0}
\expandafter\def\csname pct@rev.desc.d7.sspro.tau-T-by-type.p1.grid2-d3.icon-minus-text-on-apps-with-both.ci90.0\endcsname{5.8}
\expandafter\def\csname ppt@rev.desc.d7.sspro.tau-T-by-type.p1.grid2-d3.icon-minus-text-on-apps-with-both.ci90.0\endcsname{$+$5.8}
\expandafter\def\csname val@rev.desc.d7.sspro.tau-T-by-type.p1.grid2-d3.icon-minus-text-on-apps-with-both.ci90.1\endcsname{0.543}
\expandafter\def\csname cnt@rev.desc.d7.sspro.tau-T-by-type.p1.grid2-d3.icon-minus-text-on-apps-with-both.ci90.1\endcsname{1}
\expandafter\def\csname pct@rev.desc.d7.sspro.tau-T-by-type.p1.grid2-d3.icon-minus-text-on-apps-with-both.ci90.1\endcsname{54.3}
\expandafter\def\csname ppt@rev.desc.d7.sspro.tau-T-by-type.p1.grid2-d3.icon-minus-text-on-apps-with-both.ci90.1\endcsname{$+$54.3}
\expandafter\def\csname val@rev.desc.d7.sspro.tau-T-by-type.p1.grid2-d3.icon-minus-text-on-apps-with-both.ci95-excludes-0\endcsname{1}
\expandafter\def\csname cnt@rev.desc.d7.sspro.tau-T-by-type.p1.grid2-d3.icon-minus-text-on-apps-with-both.ci95-excludes-0\endcsname{1}
\expandafter\def\csname pct@rev.desc.d7.sspro.tau-T-by-type.p1.grid2-d3.icon-minus-text-on-apps-with-both.ci95-excludes-0\endcsname{100.0}
\expandafter\def\csname ppt@rev.desc.d7.sspro.tau-T-by-type.p1.grid2-d3.icon-minus-text-on-apps-with-both.ci95-excludes-0\endcsname{$+$100.0}
\expandafter\def\csname val@rev.desc.d7.sspro.tau-T-by-type.p1.grid2-d3.icon-minus-text-on-apps-with-both.ci95.0\endcsname{0.016}
\expandafter\def\csname cnt@rev.desc.d7.sspro.tau-T-by-type.p1.grid2-d3.icon-minus-text-on-apps-with-both.ci95.0\endcsname{0}
\expandafter\def\csname pct@rev.desc.d7.sspro.tau-T-by-type.p1.grid2-d3.icon-minus-text-on-apps-with-both.ci95.0\endcsname{1.6}
\expandafter\def\csname ppt@rev.desc.d7.sspro.tau-T-by-type.p1.grid2-d3.icon-minus-text-on-apps-with-both.ci95.0\endcsname{$+$1.6}
\expandafter\def\csname val@rev.desc.d7.sspro.tau-T-by-type.p1.grid2-d3.icon-minus-text-on-apps-with-both.ci95.1\endcsname{0.595}
\expandafter\def\csname cnt@rev.desc.d7.sspro.tau-T-by-type.p1.grid2-d3.icon-minus-text-on-apps-with-both.ci95.1\endcsname{1}
\expandafter\def\csname pct@rev.desc.d7.sspro.tau-T-by-type.p1.grid2-d3.icon-minus-text-on-apps-with-both.ci95.1\endcsname{59.5}
\expandafter\def\csname ppt@rev.desc.d7.sspro.tau-T-by-type.p1.grid2-d3.icon-minus-text-on-apps-with-both.ci95.1\endcsname{$+$59.5}
\expandafter\def\csname val@rev.desc.d7.sspro.tau-T-by-type.p1.grid2-d3.icon-minus-text-on-apps-with-both.mean\endcsname{0.289}
\expandafter\def\csname cnt@rev.desc.d7.sspro.tau-T-by-type.p1.grid2-d3.icon-minus-text-on-apps-with-both.mean\endcsname{0}
\expandafter\def\csname pct@rev.desc.d7.sspro.tau-T-by-type.p1.grid2-d3.icon-minus-text-on-apps-with-both.mean\endcsname{28.9}
\expandafter\def\csname ppt@rev.desc.d7.sspro.tau-T-by-type.p1.grid2-d3.icon-minus-text-on-apps-with-both.mean\endcsname{$+$28.9}
\expandafter\def\csname val@rev.desc.d7.sspro.tau-T-by-type.p1.grid2-d3.icon-minus-text-on-apps-with-both.n-apps\endcsname{10}
\expandafter\def\csname cnt@rev.desc.d7.sspro.tau-T-by-type.p1.grid2-d3.icon-minus-text-on-apps-with-both.n-apps\endcsname{10}
\expandafter\def\csname pct@rev.desc.d7.sspro.tau-T-by-type.p1.grid2-d3.icon-minus-text-on-apps-with-both.n-apps\endcsname{1000.0}
\expandafter\def\csname ppt@rev.desc.d7.sspro.tau-T-by-type.p1.grid2-d3.icon-minus-text-on-apps-with-both.n-apps\endcsname{$+$1000.0}
\expandafter\def\csname val@rev.desc.d7.sspro.tau-T-by-type.p1.grid2-d3.icon-minus-text-on-apps-with-both.tost1.ci90-inside\endcsname{0}
\expandafter\def\csname cnt@rev.desc.d7.sspro.tau-T-by-type.p1.grid2-d3.icon-minus-text-on-apps-with-both.tost1.ci90-inside\endcsname{0}
\expandafter\def\csname pct@rev.desc.d7.sspro.tau-T-by-type.p1.grid2-d3.icon-minus-text-on-apps-with-both.tost1.ci90-inside\endcsname{0.0}
\expandafter\def\csname ppt@rev.desc.d7.sspro.tau-T-by-type.p1.grid2-d3.icon-minus-text-on-apps-with-both.tost1.ci90-inside\endcsname{0.0}
\expandafter\def\csname val@rev.desc.d7.sspro.tau-T-by-type.p1.grid2-d3.icon-minus-text-on-apps-with-both.tost1.equivalent\endcsname{0}
\expandafter\def\csname cnt@rev.desc.d7.sspro.tau-T-by-type.p1.grid2-d3.icon-minus-text-on-apps-with-both.tost1.equivalent\endcsname{0}
\expandafter\def\csname pct@rev.desc.d7.sspro.tau-T-by-type.p1.grid2-d3.icon-minus-text-on-apps-with-both.tost1.equivalent\endcsname{0.0}
\expandafter\def\csname ppt@rev.desc.d7.sspro.tau-T-by-type.p1.grid2-d3.icon-minus-text-on-apps-with-both.tost1.equivalent\endcsname{0.0}
\expandafter\def\csname val@rev.desc.d7.sspro.tau-T-by-type.p1.grid2-d3.icon-minus-text-on-apps-with-both.tost1.p\endcsname{0.978}
\expandafter\def\csname cnt@rev.desc.d7.sspro.tau-T-by-type.p1.grid2-d3.icon-minus-text-on-apps-with-both.tost1.p\endcsname{1}
\expandafter\def\csname pct@rev.desc.d7.sspro.tau-T-by-type.p1.grid2-d3.icon-minus-text-on-apps-with-both.tost1.p\endcsname{97.8}
\expandafter\def\csname ppt@rev.desc.d7.sspro.tau-T-by-type.p1.grid2-d3.icon-minus-text-on-apps-with-both.tost1.p\endcsname{$+$97.8}
\expandafter\def\csname val@rev.desc.d7.sspro.tau-T-by-type.p1.grid2-d3.icon-minus-text-on-apps-with-both.tost5.ci90-inside\endcsname{0}
\expandafter\def\csname cnt@rev.desc.d7.sspro.tau-T-by-type.p1.grid2-d3.icon-minus-text-on-apps-with-both.tost5.ci90-inside\endcsname{0}
\expandafter\def\csname pct@rev.desc.d7.sspro.tau-T-by-type.p1.grid2-d3.icon-minus-text-on-apps-with-both.tost5.ci90-inside\endcsname{0.0}
\expandafter\def\csname ppt@rev.desc.d7.sspro.tau-T-by-type.p1.grid2-d3.icon-minus-text-on-apps-with-both.tost5.ci90-inside\endcsname{0.0}
\expandafter\def\csname val@rev.desc.d7.sspro.tau-T-by-type.p1.grid2-d3.icon-minus-text-on-apps-with-both.tost5.equivalent\endcsname{0}
\expandafter\def\csname cnt@rev.desc.d7.sspro.tau-T-by-type.p1.grid2-d3.icon-minus-text-on-apps-with-both.tost5.equivalent\endcsname{0}
\expandafter\def\csname pct@rev.desc.d7.sspro.tau-T-by-type.p1.grid2-d3.icon-minus-text-on-apps-with-both.tost5.equivalent\endcsname{0.0}
\expandafter\def\csname ppt@rev.desc.d7.sspro.tau-T-by-type.p1.grid2-d3.icon-minus-text-on-apps-with-both.tost5.equivalent\endcsname{0.0}
\expandafter\def\csname val@rev.desc.d7.sspro.tau-T-by-type.p1.grid2-d3.icon-minus-text-on-apps-with-both.tost5.p\endcsname{0.958}
\expandafter\def\csname cnt@rev.desc.d7.sspro.tau-T-by-type.p1.grid2-d3.icon-minus-text-on-apps-with-both.tost5.p\endcsname{1}
\expandafter\def\csname pct@rev.desc.d7.sspro.tau-T-by-type.p1.grid2-d3.icon-minus-text-on-apps-with-both.tost5.p\endcsname{95.8}
\expandafter\def\csname ppt@rev.desc.d7.sspro.tau-T-by-type.p1.grid2-d3.icon-minus-text-on-apps-with-both.tost5.p\endcsname{$+$95.8}
\expandafter\def\csname val@rev.desc.d7.sspro.tau-T-by-type.p1.grid2-d3.icon.acc-F\endcsname{0.786}
\expandafter\def\csname cnt@rev.desc.d7.sspro.tau-T-by-type.p1.grid2-d3.icon.acc-F\endcsname{1}
\expandafter\def\csname pct@rev.desc.d7.sspro.tau-T-by-type.p1.grid2-d3.icon.acc-F\endcsname{78.6}
\expandafter\def\csname ppt@rev.desc.d7.sspro.tau-T-by-type.p1.grid2-d3.icon.acc-F\endcsname{$+$78.6}
\expandafter\def\csname val@rev.desc.d7.sspro.tau-T-by-type.p1.grid2-d3.icon.acc-direct\endcsname{0.643}
\expandafter\def\csname cnt@rev.desc.d7.sspro.tau-T-by-type.p1.grid2-d3.icon.acc-direct\endcsname{1}
\expandafter\def\csname pct@rev.desc.d7.sspro.tau-T-by-type.p1.grid2-d3.icon.acc-direct\endcsname{64.3}
\expandafter\def\csname ppt@rev.desc.d7.sspro.tau-T-by-type.p1.grid2-d3.icon.acc-direct\endcsname{$+$64.3}
\expandafter\def\csname val@rev.desc.d7.sspro.tau-T-by-type.p1.grid2-d3.icon.n-apps\endcsname{10}
\expandafter\def\csname cnt@rev.desc.d7.sspro.tau-T-by-type.p1.grid2-d3.icon.n-apps\endcsname{10}
\expandafter\def\csname pct@rev.desc.d7.sspro.tau-T-by-type.p1.grid2-d3.icon.n-apps\endcsname{1000.0}
\expandafter\def\csname ppt@rev.desc.d7.sspro.tau-T-by-type.p1.grid2-d3.icon.n-apps\endcsname{$+$1000.0}
\expandafter\def\csname val@rev.desc.d7.sspro.tau-T-by-type.p1.grid2-d3.icon.n-tasks\endcsname{28}
\expandafter\def\csname cnt@rev.desc.d7.sspro.tau-T-by-type.p1.grid2-d3.icon.n-tasks\endcsname{28}
\expandafter\def\csname pct@rev.desc.d7.sspro.tau-T-by-type.p1.grid2-d3.icon.n-tasks\endcsname{2800.0}
\expandafter\def\csname ppt@rev.desc.d7.sspro.tau-T-by-type.p1.grid2-d3.icon.n-tasks\endcsname{$+$2800.0}
\expandafter\def\csname val@rev.desc.d7.sspro.tau-T-by-type.p1.grid2-d3.icon.tau.ci90.0\endcsname{0.050}
\expandafter\def\csname cnt@rev.desc.d7.sspro.tau-T-by-type.p1.grid2-d3.icon.tau.ci90.0\endcsname{0}
\expandafter\def\csname pct@rev.desc.d7.sspro.tau-T-by-type.p1.grid2-d3.icon.tau.ci90.0\endcsname{5.0}
\expandafter\def\csname ppt@rev.desc.d7.sspro.tau-T-by-type.p1.grid2-d3.icon.tau.ci90.0\endcsname{$+$5.0}
\expandafter\def\csname val@rev.desc.d7.sspro.tau-T-by-type.p1.grid2-d3.icon.tau.ci90.1\endcsname{0.480}
\expandafter\def\csname cnt@rev.desc.d7.sspro.tau-T-by-type.p1.grid2-d3.icon.tau.ci90.1\endcsname{0}
\expandafter\def\csname pct@rev.desc.d7.sspro.tau-T-by-type.p1.grid2-d3.icon.tau.ci90.1\endcsname{48.0}
\expandafter\def\csname ppt@rev.desc.d7.sspro.tau-T-by-type.p1.grid2-d3.icon.tau.ci90.1\endcsname{$+$48.0}
\expandafter\def\csname val@rev.desc.d7.sspro.tau-T-by-type.p1.grid2-d3.icon.tau.ci95-excludes-0\endcsname{1}
\expandafter\def\csname cnt@rev.desc.d7.sspro.tau-T-by-type.p1.grid2-d3.icon.tau.ci95-excludes-0\endcsname{1}
\expandafter\def\csname pct@rev.desc.d7.sspro.tau-T-by-type.p1.grid2-d3.icon.tau.ci95-excludes-0\endcsname{100.0}
\expandafter\def\csname ppt@rev.desc.d7.sspro.tau-T-by-type.p1.grid2-d3.icon.tau.ci95-excludes-0\endcsname{$+$100.0}
\expandafter\def\csname val@rev.desc.d7.sspro.tau-T-by-type.p1.grid2-d3.icon.tau.ci95.0\endcsname{0.020}
\expandafter\def\csname cnt@rev.desc.d7.sspro.tau-T-by-type.p1.grid2-d3.icon.tau.ci95.0\endcsname{0}
\expandafter\def\csname pct@rev.desc.d7.sspro.tau-T-by-type.p1.grid2-d3.icon.tau.ci95.0\endcsname{2.0}
\expandafter\def\csname ppt@rev.desc.d7.sspro.tau-T-by-type.p1.grid2-d3.icon.tau.ci95.0\endcsname{$+$2.0}
\expandafter\def\csname val@rev.desc.d7.sspro.tau-T-by-type.p1.grid2-d3.icon.tau.ci95.1\endcsname{0.520}
\expandafter\def\csname cnt@rev.desc.d7.sspro.tau-T-by-type.p1.grid2-d3.icon.tau.ci95.1\endcsname{1}
\expandafter\def\csname pct@rev.desc.d7.sspro.tau-T-by-type.p1.grid2-d3.icon.tau.ci95.1\endcsname{52.0}
\expandafter\def\csname ppt@rev.desc.d7.sspro.tau-T-by-type.p1.grid2-d3.icon.tau.ci95.1\endcsname{$+$52.0}
\expandafter\def\csname val@rev.desc.d7.sspro.tau-T-by-type.p1.grid2-d3.icon.tau.mean\endcsname{0.250}
\expandafter\def\csname cnt@rev.desc.d7.sspro.tau-T-by-type.p1.grid2-d3.icon.tau.mean\endcsname{0}
\expandafter\def\csname pct@rev.desc.d7.sspro.tau-T-by-type.p1.grid2-d3.icon.tau.mean\endcsname{25.0}
\expandafter\def\csname ppt@rev.desc.d7.sspro.tau-T-by-type.p1.grid2-d3.icon.tau.mean\endcsname{$+$25.0}
\expandafter\def\csname val@rev.desc.d7.sspro.tau-T-by-type.p1.grid2-d3.icon.tau.n-apps\endcsname{10}
\expandafter\def\csname cnt@rev.desc.d7.sspro.tau-T-by-type.p1.grid2-d3.icon.tau.n-apps\endcsname{10}
\expandafter\def\csname pct@rev.desc.d7.sspro.tau-T-by-type.p1.grid2-d3.icon.tau.n-apps\endcsname{1000.0}
\expandafter\def\csname ppt@rev.desc.d7.sspro.tau-T-by-type.p1.grid2-d3.icon.tau.n-apps\endcsname{$+$1000.0}
\expandafter\def\csname val@rev.desc.d7.sspro.tau-T-by-type.p1.grid2-d3.icon.tau.tost1.ci90-inside\endcsname{0}
\expandafter\def\csname cnt@rev.desc.d7.sspro.tau-T-by-type.p1.grid2-d3.icon.tau.tost1.ci90-inside\endcsname{0}
\expandafter\def\csname pct@rev.desc.d7.sspro.tau-T-by-type.p1.grid2-d3.icon.tau.tost1.ci90-inside\endcsname{0.0}
\expandafter\def\csname ppt@rev.desc.d7.sspro.tau-T-by-type.p1.grid2-d3.icon.tau.tost1.ci90-inside\endcsname{0.0}
\expandafter\def\csname val@rev.desc.d7.sspro.tau-T-by-type.p1.grid2-d3.icon.tau.tost1.equivalent\endcsname{0}
\expandafter\def\csname cnt@rev.desc.d7.sspro.tau-T-by-type.p1.grid2-d3.icon.tau.tost1.equivalent\endcsname{0}
\expandafter\def\csname pct@rev.desc.d7.sspro.tau-T-by-type.p1.grid2-d3.icon.tau.tost1.equivalent\endcsname{0.0}
\expandafter\def\csname ppt@rev.desc.d7.sspro.tau-T-by-type.p1.grid2-d3.icon.tau.tost1.equivalent\endcsname{0.0}
\expandafter\def\csname val@rev.desc.d7.sspro.tau-T-by-type.p1.grid2-d3.icon.tau.tost1.p\endcsname{0.978}
\expandafter\def\csname cnt@rev.desc.d7.sspro.tau-T-by-type.p1.grid2-d3.icon.tau.tost1.p\endcsname{1}
\expandafter\def\csname pct@rev.desc.d7.sspro.tau-T-by-type.p1.grid2-d3.icon.tau.tost1.p\endcsname{97.8}
\expandafter\def\csname ppt@rev.desc.d7.sspro.tau-T-by-type.p1.grid2-d3.icon.tau.tost1.p\endcsname{$+$97.8}
\expandafter\def\csname val@rev.desc.d7.sspro.tau-T-by-type.p1.grid2-d3.icon.tau.tost5.ci90-inside\endcsname{0}
\expandafter\def\csname cnt@rev.desc.d7.sspro.tau-T-by-type.p1.grid2-d3.icon.tau.tost5.ci90-inside\endcsname{0}
\expandafter\def\csname pct@rev.desc.d7.sspro.tau-T-by-type.p1.grid2-d3.icon.tau.tost5.ci90-inside\endcsname{0.0}
\expandafter\def\csname ppt@rev.desc.d7.sspro.tau-T-by-type.p1.grid2-d3.icon.tau.tost5.ci90-inside\endcsname{0.0}
\expandafter\def\csname val@rev.desc.d7.sspro.tau-T-by-type.p1.grid2-d3.icon.tau.tost5.equivalent\endcsname{0}
\expandafter\def\csname cnt@rev.desc.d7.sspro.tau-T-by-type.p1.grid2-d3.icon.tau.tost5.equivalent\endcsname{0}
\expandafter\def\csname pct@rev.desc.d7.sspro.tau-T-by-type.p1.grid2-d3.icon.tau.tost5.equivalent\endcsname{0.0}
\expandafter\def\csname ppt@rev.desc.d7.sspro.tau-T-by-type.p1.grid2-d3.icon.tau.tost5.equivalent\endcsname{0.0}
\expandafter\def\csname val@rev.desc.d7.sspro.tau-T-by-type.p1.grid2-d3.icon.tau.tost5.p\endcsname{0.949}
\expandafter\def\csname cnt@rev.desc.d7.sspro.tau-T-by-type.p1.grid2-d3.icon.tau.tost5.p\endcsname{1}
\expandafter\def\csname pct@rev.desc.d7.sspro.tau-T-by-type.p1.grid2-d3.icon.tau.tost5.p\endcsname{94.9}
\expandafter\def\csname ppt@rev.desc.d7.sspro.tau-T-by-type.p1.grid2-d3.icon.tau.tost5.p\endcsname{$+$94.9}
\expandafter\def\csname val@rev.desc.d7.sspro.tau-T-by-type.p1.grid2-d3.text.acc-F\endcsname{0.882}
\expandafter\def\csname cnt@rev.desc.d7.sspro.tau-T-by-type.p1.grid2-d3.text.acc-F\endcsname{1}
\expandafter\def\csname pct@rev.desc.d7.sspro.tau-T-by-type.p1.grid2-d3.text.acc-F\endcsname{88.2}
\expandafter\def\csname ppt@rev.desc.d7.sspro.tau-T-by-type.p1.grid2-d3.text.acc-F\endcsname{$+$88.2}
\expandafter\def\csname val@rev.desc.d7.sspro.tau-T-by-type.p1.grid2-d3.text.acc-direct\endcsname{0.779}
\expandafter\def\csname cnt@rev.desc.d7.sspro.tau-T-by-type.p1.grid2-d3.text.acc-direct\endcsname{1}
\expandafter\def\csname pct@rev.desc.d7.sspro.tau-T-by-type.p1.grid2-d3.text.acc-direct\endcsname{77.9}
\expandafter\def\csname ppt@rev.desc.d7.sspro.tau-T-by-type.p1.grid2-d3.text.acc-direct\endcsname{$+$77.9}
\expandafter\def\csname val@rev.desc.d7.sspro.tau-T-by-type.p1.grid2-d3.text.n-apps\endcsname{12}
\expandafter\def\csname cnt@rev.desc.d7.sspro.tau-T-by-type.p1.grid2-d3.text.n-apps\endcsname{12}
\expandafter\def\csname pct@rev.desc.d7.sspro.tau-T-by-type.p1.grid2-d3.text.n-apps\endcsname{1200.0}
\expandafter\def\csname ppt@rev.desc.d7.sspro.tau-T-by-type.p1.grid2-d3.text.n-apps\endcsname{$+$1200.0}
\expandafter\def\csname val@rev.desc.d7.sspro.tau-T-by-type.p1.grid2-d3.text.n-tasks\endcsname{68}
\expandafter\def\csname cnt@rev.desc.d7.sspro.tau-T-by-type.p1.grid2-d3.text.n-tasks\endcsname{68}
\expandafter\def\csname pct@rev.desc.d7.sspro.tau-T-by-type.p1.grid2-d3.text.n-tasks\endcsname{6800.0}
\expandafter\def\csname ppt@rev.desc.d7.sspro.tau-T-by-type.p1.grid2-d3.text.n-tasks\endcsname{$+$6800.0}
\expandafter\def\csname val@rev.desc.d7.sspro.tau-T-by-type.p1.grid2-d3.text.tau.ci90.0\endcsname{-0.061}
\expandafter\def\csname cnt@rev.desc.d7.sspro.tau-T-by-type.p1.grid2-d3.text.tau.ci90.0\endcsname{0}
\expandafter\def\csname pct@rev.desc.d7.sspro.tau-T-by-type.p1.grid2-d3.text.tau.ci90.0\endcsname{$-$6.1}
\expandafter\def\csname ppt@rev.desc.d7.sspro.tau-T-by-type.p1.grid2-d3.text.tau.ci90.0\endcsname{$-$6.1}
\expandafter\def\csname val@rev.desc.d7.sspro.tau-T-by-type.p1.grid2-d3.text.tau.ci90.1\endcsname{0.169}
\expandafter\def\csname cnt@rev.desc.d7.sspro.tau-T-by-type.p1.grid2-d3.text.tau.ci90.1\endcsname{0}
\expandafter\def\csname pct@rev.desc.d7.sspro.tau-T-by-type.p1.grid2-d3.text.tau.ci90.1\endcsname{16.9}
\expandafter\def\csname ppt@rev.desc.d7.sspro.tau-T-by-type.p1.grid2-d3.text.tau.ci90.1\endcsname{$+$16.9}
\expandafter\def\csname val@rev.desc.d7.sspro.tau-T-by-type.p1.grid2-d3.text.tau.ci95-excludes-0\endcsname{0}
\expandafter\def\csname cnt@rev.desc.d7.sspro.tau-T-by-type.p1.grid2-d3.text.tau.ci95-excludes-0\endcsname{0}
\expandafter\def\csname pct@rev.desc.d7.sspro.tau-T-by-type.p1.grid2-d3.text.tau.ci95-excludes-0\endcsname{0.0}
\expandafter\def\csname ppt@rev.desc.d7.sspro.tau-T-by-type.p1.grid2-d3.text.tau.ci95-excludes-0\endcsname{0.0}
\expandafter\def\csname val@rev.desc.d7.sspro.tau-T-by-type.p1.grid2-d3.text.tau.ci95.0\endcsname{-0.079}
\expandafter\def\csname cnt@rev.desc.d7.sspro.tau-T-by-type.p1.grid2-d3.text.tau.ci95.0\endcsname{0}
\expandafter\def\csname pct@rev.desc.d7.sspro.tau-T-by-type.p1.grid2-d3.text.tau.ci95.0\endcsname{$-$7.9}
\expandafter\def\csname ppt@rev.desc.d7.sspro.tau-T-by-type.p1.grid2-d3.text.tau.ci95.0\endcsname{$-$7.9}
\expandafter\def\csname val@rev.desc.d7.sspro.tau-T-by-type.p1.grid2-d3.text.tau.ci95.1\endcsname{0.194}
\expandafter\def\csname cnt@rev.desc.d7.sspro.tau-T-by-type.p1.grid2-d3.text.tau.ci95.1\endcsname{0}
\expandafter\def\csname pct@rev.desc.d7.sspro.tau-T-by-type.p1.grid2-d3.text.tau.ci95.1\endcsname{19.4}
\expandafter\def\csname ppt@rev.desc.d7.sspro.tau-T-by-type.p1.grid2-d3.text.tau.ci95.1\endcsname{$+$19.4}
\expandafter\def\csname val@rev.desc.d7.sspro.tau-T-by-type.p1.grid2-d3.text.tau.mean\endcsname{0.051}
\expandafter\def\csname cnt@rev.desc.d7.sspro.tau-T-by-type.p1.grid2-d3.text.tau.mean\endcsname{0}
\expandafter\def\csname pct@rev.desc.d7.sspro.tau-T-by-type.p1.grid2-d3.text.tau.mean\endcsname{5.1}
\expandafter\def\csname ppt@rev.desc.d7.sspro.tau-T-by-type.p1.grid2-d3.text.tau.mean\endcsname{$+$5.1}
\expandafter\def\csname val@rev.desc.d7.sspro.tau-T-by-type.p1.grid2-d3.text.tau.n-apps\endcsname{12}
\expandafter\def\csname cnt@rev.desc.d7.sspro.tau-T-by-type.p1.grid2-d3.text.tau.n-apps\endcsname{12}
\expandafter\def\csname pct@rev.desc.d7.sspro.tau-T-by-type.p1.grid2-d3.text.tau.n-apps\endcsname{1200.0}
\expandafter\def\csname ppt@rev.desc.d7.sspro.tau-T-by-type.p1.grid2-d3.text.tau.n-apps\endcsname{$+$1200.0}
\expandafter\def\csname val@rev.desc.d7.sspro.tau-T-by-type.p1.grid2-d3.text.tau.tost1.ci90-inside\endcsname{0}
\expandafter\def\csname cnt@rev.desc.d7.sspro.tau-T-by-type.p1.grid2-d3.text.tau.tost1.ci90-inside\endcsname{0}
\expandafter\def\csname pct@rev.desc.d7.sspro.tau-T-by-type.p1.grid2-d3.text.tau.tost1.ci90-inside\endcsname{0.0}
\expandafter\def\csname ppt@rev.desc.d7.sspro.tau-T-by-type.p1.grid2-d3.text.tau.tost1.ci90-inside\endcsname{0.0}
\expandafter\def\csname val@rev.desc.d7.sspro.tau-T-by-type.p1.grid2-d3.text.tau.tost1.equivalent\endcsname{0}
\expandafter\def\csname cnt@rev.desc.d7.sspro.tau-T-by-type.p1.grid2-d3.text.tau.tost1.equivalent\endcsname{0}
\expandafter\def\csname pct@rev.desc.d7.sspro.tau-T-by-type.p1.grid2-d3.text.tau.tost1.equivalent\endcsname{0.0}
\expandafter\def\csname ppt@rev.desc.d7.sspro.tau-T-by-type.p1.grid2-d3.text.tau.tost1.equivalent\endcsname{0.0}
\expandafter\def\csname val@rev.desc.d7.sspro.tau-T-by-type.p1.grid2-d3.text.tau.tost1.p\endcsname{0.709}
\expandafter\def\csname cnt@rev.desc.d7.sspro.tau-T-by-type.p1.grid2-d3.text.tau.tost1.p\endcsname{1}
\expandafter\def\csname pct@rev.desc.d7.sspro.tau-T-by-type.p1.grid2-d3.text.tau.tost1.p\endcsname{70.9}
\expandafter\def\csname ppt@rev.desc.d7.sspro.tau-T-by-type.p1.grid2-d3.text.tau.tost1.p\endcsname{$+$70.9}
\expandafter\def\csname val@rev.desc.d7.sspro.tau-T-by-type.p1.grid2-d3.text.tau.tost5.ci90-inside\endcsname{0}
\expandafter\def\csname cnt@rev.desc.d7.sspro.tau-T-by-type.p1.grid2-d3.text.tau.tost5.ci90-inside\endcsname{0}
\expandafter\def\csname pct@rev.desc.d7.sspro.tau-T-by-type.p1.grid2-d3.text.tau.tost5.ci90-inside\endcsname{0.0}
\expandafter\def\csname ppt@rev.desc.d7.sspro.tau-T-by-type.p1.grid2-d3.text.tau.tost5.ci90-inside\endcsname{0.0}
\expandafter\def\csname val@rev.desc.d7.sspro.tau-T-by-type.p1.grid2-d3.text.tau.tost5.equivalent\endcsname{0}
\expandafter\def\csname cnt@rev.desc.d7.sspro.tau-T-by-type.p1.grid2-d3.text.tau.tost5.equivalent\endcsname{0}
\expandafter\def\csname pct@rev.desc.d7.sspro.tau-T-by-type.p1.grid2-d3.text.tau.tost5.equivalent\endcsname{0.0}
\expandafter\def\csname ppt@rev.desc.d7.sspro.tau-T-by-type.p1.grid2-d3.text.tau.tost5.equivalent\endcsname{0.0}
\expandafter\def\csname val@rev.desc.d7.sspro.tau-T-by-type.p1.grid2-d3.text.tau.tost5.p\endcsname{0.501}
\expandafter\def\csname cnt@rev.desc.d7.sspro.tau-T-by-type.p1.grid2-d3.text.tau.tost5.p\endcsname{1}
\expandafter\def\csname pct@rev.desc.d7.sspro.tau-T-by-type.p1.grid2-d3.text.tau.tost5.p\endcsname{50.1}
\expandafter\def\csname ppt@rev.desc.d7.sspro.tau-T-by-type.p1.grid2-d3.text.tau.tost5.p\endcsname{$+$50.1}
\expandafter\def\csname val@rev.desc.d7.sspro.tau-T-by-type.p1.zoomclick-d3.icon-minus-text-on-apps-with-both.ci90.0\endcsname{0.005}
\expandafter\def\csname cnt@rev.desc.d7.sspro.tau-T-by-type.p1.zoomclick-d3.icon-minus-text-on-apps-with-both.ci90.0\endcsname{0}
\expandafter\def\csname pct@rev.desc.d7.sspro.tau-T-by-type.p1.zoomclick-d3.icon-minus-text-on-apps-with-both.ci90.0\endcsname{0.5}
\expandafter\def\csname ppt@rev.desc.d7.sspro.tau-T-by-type.p1.zoomclick-d3.icon-minus-text-on-apps-with-both.ci90.0\endcsname{$+$0.5}
\expandafter\def\csname val@rev.desc.d7.sspro.tau-T-by-type.p1.zoomclick-d3.icon-minus-text-on-apps-with-both.ci90.1\endcsname{0.383}
\expandafter\def\csname cnt@rev.desc.d7.sspro.tau-T-by-type.p1.zoomclick-d3.icon-minus-text-on-apps-with-both.ci90.1\endcsname{0}
\expandafter\def\csname pct@rev.desc.d7.sspro.tau-T-by-type.p1.zoomclick-d3.icon-minus-text-on-apps-with-both.ci90.1\endcsname{38.3}
\expandafter\def\csname ppt@rev.desc.d7.sspro.tau-T-by-type.p1.zoomclick-d3.icon-minus-text-on-apps-with-both.ci90.1\endcsname{$+$38.3}
\expandafter\def\csname val@rev.desc.d7.sspro.tau-T-by-type.p1.zoomclick-d3.icon-minus-text-on-apps-with-both.ci95-excludes-0\endcsname{0}
\expandafter\def\csname cnt@rev.desc.d7.sspro.tau-T-by-type.p1.zoomclick-d3.icon-minus-text-on-apps-with-both.ci95-excludes-0\endcsname{0}
\expandafter\def\csname pct@rev.desc.d7.sspro.tau-T-by-type.p1.zoomclick-d3.icon-minus-text-on-apps-with-both.ci95-excludes-0\endcsname{0.0}
\expandafter\def\csname ppt@rev.desc.d7.sspro.tau-T-by-type.p1.zoomclick-d3.icon-minus-text-on-apps-with-both.ci95-excludes-0\endcsname{0.0}
\expandafter\def\csname val@rev.desc.d7.sspro.tau-T-by-type.p1.zoomclick-d3.icon-minus-text-on-apps-with-both.ci95.0\endcsname{-0.024}
\expandafter\def\csname cnt@rev.desc.d7.sspro.tau-T-by-type.p1.zoomclick-d3.icon-minus-text-on-apps-with-both.ci95.0\endcsname{0}
\expandafter\def\csname pct@rev.desc.d7.sspro.tau-T-by-type.p1.zoomclick-d3.icon-minus-text-on-apps-with-both.ci95.0\endcsname{$-$2.4}
\expandafter\def\csname ppt@rev.desc.d7.sspro.tau-T-by-type.p1.zoomclick-d3.icon-minus-text-on-apps-with-both.ci95.0\endcsname{$-$2.4}
\expandafter\def\csname val@rev.desc.d7.sspro.tau-T-by-type.p1.zoomclick-d3.icon-minus-text-on-apps-with-both.ci95.1\endcsname{0.423}
\expandafter\def\csname cnt@rev.desc.d7.sspro.tau-T-by-type.p1.zoomclick-d3.icon-minus-text-on-apps-with-both.ci95.1\endcsname{0}
\expandafter\def\csname pct@rev.desc.d7.sspro.tau-T-by-type.p1.zoomclick-d3.icon-minus-text-on-apps-with-both.ci95.1\endcsname{42.3}
\expandafter\def\csname ppt@rev.desc.d7.sspro.tau-T-by-type.p1.zoomclick-d3.icon-minus-text-on-apps-with-both.ci95.1\endcsname{$+$42.3}
\expandafter\def\csname val@rev.desc.d7.sspro.tau-T-by-type.p1.zoomclick-d3.icon-minus-text-on-apps-with-both.mean\endcsname{0.188}
\expandafter\def\csname cnt@rev.desc.d7.sspro.tau-T-by-type.p1.zoomclick-d3.icon-minus-text-on-apps-with-both.mean\endcsname{0}
\expandafter\def\csname pct@rev.desc.d7.sspro.tau-T-by-type.p1.zoomclick-d3.icon-minus-text-on-apps-with-both.mean\endcsname{18.8}
\expandafter\def\csname ppt@rev.desc.d7.sspro.tau-T-by-type.p1.zoomclick-d3.icon-minus-text-on-apps-with-both.mean\endcsname{$+$18.8}
\expandafter\def\csname val@rev.desc.d7.sspro.tau-T-by-type.p1.zoomclick-d3.icon-minus-text-on-apps-with-both.n-apps\endcsname{10}
\expandafter\def\csname cnt@rev.desc.d7.sspro.tau-T-by-type.p1.zoomclick-d3.icon-minus-text-on-apps-with-both.n-apps\endcsname{10}
\expandafter\def\csname pct@rev.desc.d7.sspro.tau-T-by-type.p1.zoomclick-d3.icon-minus-text-on-apps-with-both.n-apps\endcsname{1000.0}
\expandafter\def\csname ppt@rev.desc.d7.sspro.tau-T-by-type.p1.zoomclick-d3.icon-minus-text-on-apps-with-both.n-apps\endcsname{$+$1000.0}
\expandafter\def\csname val@rev.desc.d7.sspro.tau-T-by-type.p1.zoomclick-d3.icon-minus-text-on-apps-with-both.tost1.ci90-inside\endcsname{0}
\expandafter\def\csname cnt@rev.desc.d7.sspro.tau-T-by-type.p1.zoomclick-d3.icon-minus-text-on-apps-with-both.tost1.ci90-inside\endcsname{0}
\expandafter\def\csname pct@rev.desc.d7.sspro.tau-T-by-type.p1.zoomclick-d3.icon-minus-text-on-apps-with-both.tost1.ci90-inside\endcsname{0.0}
\expandafter\def\csname ppt@rev.desc.d7.sspro.tau-T-by-type.p1.zoomclick-d3.icon-minus-text-on-apps-with-both.tost1.ci90-inside\endcsname{0.0}
\expandafter\def\csname val@rev.desc.d7.sspro.tau-T-by-type.p1.zoomclick-d3.icon-minus-text-on-apps-with-both.tost1.equivalent\endcsname{0}
\expandafter\def\csname cnt@rev.desc.d7.sspro.tau-T-by-type.p1.zoomclick-d3.icon-minus-text-on-apps-with-both.tost1.equivalent\endcsname{0}
\expandafter\def\csname pct@rev.desc.d7.sspro.tau-T-by-type.p1.zoomclick-d3.icon-minus-text-on-apps-with-both.tost1.equivalent\endcsname{0.0}
\expandafter\def\csname ppt@rev.desc.d7.sspro.tau-T-by-type.p1.zoomclick-d3.icon-minus-text-on-apps-with-both.tost1.equivalent\endcsname{0.0}
\expandafter\def\csname val@rev.desc.d7.sspro.tau-T-by-type.p1.zoomclick-d3.icon-minus-text-on-apps-with-both.tost1.p\endcsname{0.946}
\expandafter\def\csname cnt@rev.desc.d7.sspro.tau-T-by-type.p1.zoomclick-d3.icon-minus-text-on-apps-with-both.tost1.p\endcsname{1}
\expandafter\def\csname pct@rev.desc.d7.sspro.tau-T-by-type.p1.zoomclick-d3.icon-minus-text-on-apps-with-both.tost1.p\endcsname{94.6}
\expandafter\def\csname ppt@rev.desc.d7.sspro.tau-T-by-type.p1.zoomclick-d3.icon-minus-text-on-apps-with-both.tost1.p\endcsname{$+$94.6}
\expandafter\def\csname val@rev.desc.d7.sspro.tau-T-by-type.p1.zoomclick-d3.icon-minus-text-on-apps-with-both.tost5.ci90-inside\endcsname{0}
\expandafter\def\csname cnt@rev.desc.d7.sspro.tau-T-by-type.p1.zoomclick-d3.icon-minus-text-on-apps-with-both.tost5.ci90-inside\endcsname{0}
\expandafter\def\csname pct@rev.desc.d7.sspro.tau-T-by-type.p1.zoomclick-d3.icon-minus-text-on-apps-with-both.tost5.ci90-inside\endcsname{0.0}
\expandafter\def\csname ppt@rev.desc.d7.sspro.tau-T-by-type.p1.zoomclick-d3.icon-minus-text-on-apps-with-both.tost5.ci90-inside\endcsname{0.0}
\expandafter\def\csname val@rev.desc.d7.sspro.tau-T-by-type.p1.zoomclick-d3.icon-minus-text-on-apps-with-both.tost5.equivalent\endcsname{0}
\expandafter\def\csname cnt@rev.desc.d7.sspro.tau-T-by-type.p1.zoomclick-d3.icon-minus-text-on-apps-with-both.tost5.equivalent\endcsname{0}
\expandafter\def\csname pct@rev.desc.d7.sspro.tau-T-by-type.p1.zoomclick-d3.icon-minus-text-on-apps-with-both.tost5.equivalent\endcsname{0.0}
\expandafter\def\csname ppt@rev.desc.d7.sspro.tau-T-by-type.p1.zoomclick-d3.icon-minus-text-on-apps-with-both.tost5.equivalent\endcsname{0.0}
\expandafter\def\csname val@rev.desc.d7.sspro.tau-T-by-type.p1.zoomclick-d3.icon-minus-text-on-apps-with-both.tost5.p\endcsname{0.884}
\expandafter\def\csname cnt@rev.desc.d7.sspro.tau-T-by-type.p1.zoomclick-d3.icon-minus-text-on-apps-with-both.tost5.p\endcsname{1}
\expandafter\def\csname pct@rev.desc.d7.sspro.tau-T-by-type.p1.zoomclick-d3.icon-minus-text-on-apps-with-both.tost5.p\endcsname{88.4}
\expandafter\def\csname ppt@rev.desc.d7.sspro.tau-T-by-type.p1.zoomclick-d3.icon-minus-text-on-apps-with-both.tost5.p\endcsname{$+$88.4}
\expandafter\def\csname val@rev.desc.d7.sspro.tau-T-by-type.p1.zoomclick-d3.icon.acc-F\endcsname{0.821}
\expandafter\def\csname cnt@rev.desc.d7.sspro.tau-T-by-type.p1.zoomclick-d3.icon.acc-F\endcsname{1}
\expandafter\def\csname pct@rev.desc.d7.sspro.tau-T-by-type.p1.zoomclick-d3.icon.acc-F\endcsname{82.1}
\expandafter\def\csname ppt@rev.desc.d7.sspro.tau-T-by-type.p1.zoomclick-d3.icon.acc-F\endcsname{$+$82.1}
\expandafter\def\csname val@rev.desc.d7.sspro.tau-T-by-type.p1.zoomclick-d3.icon.acc-direct\endcsname{0.643}
\expandafter\def\csname cnt@rev.desc.d7.sspro.tau-T-by-type.p1.zoomclick-d3.icon.acc-direct\endcsname{1}
\expandafter\def\csname pct@rev.desc.d7.sspro.tau-T-by-type.p1.zoomclick-d3.icon.acc-direct\endcsname{64.3}
\expandafter\def\csname ppt@rev.desc.d7.sspro.tau-T-by-type.p1.zoomclick-d3.icon.acc-direct\endcsname{$+$64.3}
\expandafter\def\csname val@rev.desc.d7.sspro.tau-T-by-type.p1.zoomclick-d3.icon.n-apps\endcsname{10}
\expandafter\def\csname cnt@rev.desc.d7.sspro.tau-T-by-type.p1.zoomclick-d3.icon.n-apps\endcsname{10}
\expandafter\def\csname pct@rev.desc.d7.sspro.tau-T-by-type.p1.zoomclick-d3.icon.n-apps\endcsname{1000.0}
\expandafter\def\csname ppt@rev.desc.d7.sspro.tau-T-by-type.p1.zoomclick-d3.icon.n-apps\endcsname{$+$1000.0}
\expandafter\def\csname val@rev.desc.d7.sspro.tau-T-by-type.p1.zoomclick-d3.icon.n-tasks\endcsname{28}
\expandafter\def\csname cnt@rev.desc.d7.sspro.tau-T-by-type.p1.zoomclick-d3.icon.n-tasks\endcsname{28}
\expandafter\def\csname pct@rev.desc.d7.sspro.tau-T-by-type.p1.zoomclick-d3.icon.n-tasks\endcsname{2800.0}
\expandafter\def\csname ppt@rev.desc.d7.sspro.tau-T-by-type.p1.zoomclick-d3.icon.n-tasks\endcsname{$+$2800.0}
\expandafter\def\csname val@rev.desc.d7.sspro.tau-T-by-type.p1.zoomclick-d3.icon.tau.ci90.0\endcsname{0.103}
\expandafter\def\csname cnt@rev.desc.d7.sspro.tau-T-by-type.p1.zoomclick-d3.icon.tau.ci90.0\endcsname{0}
\expandafter\def\csname pct@rev.desc.d7.sspro.tau-T-by-type.p1.zoomclick-d3.icon.tau.ci90.0\endcsname{10.3}
\expandafter\def\csname ppt@rev.desc.d7.sspro.tau-T-by-type.p1.zoomclick-d3.icon.tau.ci90.0\endcsname{$+$10.3}
\expandafter\def\csname val@rev.desc.d7.sspro.tau-T-by-type.p1.zoomclick-d3.icon.tau.ci90.1\endcsname{0.423}
\expandafter\def\csname cnt@rev.desc.d7.sspro.tau-T-by-type.p1.zoomclick-d3.icon.tau.ci90.1\endcsname{0}
\expandafter\def\csname pct@rev.desc.d7.sspro.tau-T-by-type.p1.zoomclick-d3.icon.tau.ci90.1\endcsname{42.3}
\expandafter\def\csname ppt@rev.desc.d7.sspro.tau-T-by-type.p1.zoomclick-d3.icon.tau.ci90.1\endcsname{$+$42.3}
\expandafter\def\csname val@rev.desc.d7.sspro.tau-T-by-type.p1.zoomclick-d3.icon.tau.ci95-excludes-0\endcsname{1}
\expandafter\def\csname cnt@rev.desc.d7.sspro.tau-T-by-type.p1.zoomclick-d3.icon.tau.ci95-excludes-0\endcsname{1}
\expandafter\def\csname pct@rev.desc.d7.sspro.tau-T-by-type.p1.zoomclick-d3.icon.tau.ci95-excludes-0\endcsname{100.0}
\expandafter\def\csname ppt@rev.desc.d7.sspro.tau-T-by-type.p1.zoomclick-d3.icon.tau.ci95-excludes-0\endcsname{$+$100.0}
\expandafter\def\csname val@rev.desc.d7.sspro.tau-T-by-type.p1.zoomclick-d3.icon.tau.ci95.0\endcsname{0.083}
\expandafter\def\csname cnt@rev.desc.d7.sspro.tau-T-by-type.p1.zoomclick-d3.icon.tau.ci95.0\endcsname{0}
\expandafter\def\csname pct@rev.desc.d7.sspro.tau-T-by-type.p1.zoomclick-d3.icon.tau.ci95.0\endcsname{8.3}
\expandafter\def\csname ppt@rev.desc.d7.sspro.tau-T-by-type.p1.zoomclick-d3.icon.tau.ci95.0\endcsname{$+$8.3}
\expandafter\def\csname val@rev.desc.d7.sspro.tau-T-by-type.p1.zoomclick-d3.icon.tau.ci95.1\endcsname{0.467}
\expandafter\def\csname cnt@rev.desc.d7.sspro.tau-T-by-type.p1.zoomclick-d3.icon.tau.ci95.1\endcsname{0}
\expandafter\def\csname pct@rev.desc.d7.sspro.tau-T-by-type.p1.zoomclick-d3.icon.tau.ci95.1\endcsname{46.7}
\expandafter\def\csname ppt@rev.desc.d7.sspro.tau-T-by-type.p1.zoomclick-d3.icon.tau.ci95.1\endcsname{$+$46.7}
\expandafter\def\csname val@rev.desc.d7.sspro.tau-T-by-type.p1.zoomclick-d3.icon.tau.mean\endcsname{0.253}
\expandafter\def\csname cnt@rev.desc.d7.sspro.tau-T-by-type.p1.zoomclick-d3.icon.tau.mean\endcsname{0}
\expandafter\def\csname pct@rev.desc.d7.sspro.tau-T-by-type.p1.zoomclick-d3.icon.tau.mean\endcsname{25.3}
\expandafter\def\csname ppt@rev.desc.d7.sspro.tau-T-by-type.p1.zoomclick-d3.icon.tau.mean\endcsname{$+$25.3}
\expandafter\def\csname val@rev.desc.d7.sspro.tau-T-by-type.p1.zoomclick-d3.icon.tau.n-apps\endcsname{10}
\expandafter\def\csname cnt@rev.desc.d7.sspro.tau-T-by-type.p1.zoomclick-d3.icon.tau.n-apps\endcsname{10}
\expandafter\def\csname pct@rev.desc.d7.sspro.tau-T-by-type.p1.zoomclick-d3.icon.tau.n-apps\endcsname{1000.0}
\expandafter\def\csname ppt@rev.desc.d7.sspro.tau-T-by-type.p1.zoomclick-d3.icon.tau.n-apps\endcsname{$+$1000.0}
\expandafter\def\csname val@rev.desc.d7.sspro.tau-T-by-type.p1.zoomclick-d3.icon.tau.tost1.ci90-inside\endcsname{0}
\expandafter\def\csname cnt@rev.desc.d7.sspro.tau-T-by-type.p1.zoomclick-d3.icon.tau.tost1.ci90-inside\endcsname{0}
\expandafter\def\csname pct@rev.desc.d7.sspro.tau-T-by-type.p1.zoomclick-d3.icon.tau.tost1.ci90-inside\endcsname{0.0}
\expandafter\def\csname ppt@rev.desc.d7.sspro.tau-T-by-type.p1.zoomclick-d3.icon.tau.tost1.ci90-inside\endcsname{0.0}
\expandafter\def\csname val@rev.desc.d7.sspro.tau-T-by-type.p1.zoomclick-d3.icon.tau.tost1.equivalent\endcsname{0}
\expandafter\def\csname cnt@rev.desc.d7.sspro.tau-T-by-type.p1.zoomclick-d3.icon.tau.tost1.equivalent\endcsname{0}
\expandafter\def\csname pct@rev.desc.d7.sspro.tau-T-by-type.p1.zoomclick-d3.icon.tau.tost1.equivalent\endcsname{0.0}
\expandafter\def\csname ppt@rev.desc.d7.sspro.tau-T-by-type.p1.zoomclick-d3.icon.tau.tost1.equivalent\endcsname{0.0}
\expandafter\def\csname val@rev.desc.d7.sspro.tau-T-by-type.p1.zoomclick-d3.icon.tau.tost1.p\endcsname{1.000}
\expandafter\def\csname cnt@rev.desc.d7.sspro.tau-T-by-type.p1.zoomclick-d3.icon.tau.tost1.p\endcsname{1}
\expandafter\def\csname pct@rev.desc.d7.sspro.tau-T-by-type.p1.zoomclick-d3.icon.tau.tost1.p\endcsname{100.0}
\expandafter\def\csname ppt@rev.desc.d7.sspro.tau-T-by-type.p1.zoomclick-d3.icon.tau.tost1.p\endcsname{$+$100.0}
\expandafter\def\csname val@rev.desc.d7.sspro.tau-T-by-type.p1.zoomclick-d3.icon.tau.tost5.ci90-inside\endcsname{0}
\expandafter\def\csname cnt@rev.desc.d7.sspro.tau-T-by-type.p1.zoomclick-d3.icon.tau.tost5.ci90-inside\endcsname{0}
\expandafter\def\csname pct@rev.desc.d7.sspro.tau-T-by-type.p1.zoomclick-d3.icon.tau.tost5.ci90-inside\endcsname{0.0}
\expandafter\def\csname ppt@rev.desc.d7.sspro.tau-T-by-type.p1.zoomclick-d3.icon.tau.tost5.ci90-inside\endcsname{0.0}
\expandafter\def\csname val@rev.desc.d7.sspro.tau-T-by-type.p1.zoomclick-d3.icon.tau.tost5.equivalent\endcsname{0}
\expandafter\def\csname cnt@rev.desc.d7.sspro.tau-T-by-type.p1.zoomclick-d3.icon.tau.tost5.equivalent\endcsname{0}
\expandafter\def\csname pct@rev.desc.d7.sspro.tau-T-by-type.p1.zoomclick-d3.icon.tau.tost5.equivalent\endcsname{0.0}
\expandafter\def\csname ppt@rev.desc.d7.sspro.tau-T-by-type.p1.zoomclick-d3.icon.tau.tost5.equivalent\endcsname{0.0}
\expandafter\def\csname val@rev.desc.d7.sspro.tau-T-by-type.p1.zoomclick-d3.icon.tau.tost5.p\endcsname{0.994}
\expandafter\def\csname cnt@rev.desc.d7.sspro.tau-T-by-type.p1.zoomclick-d3.icon.tau.tost5.p\endcsname{1}
\expandafter\def\csname pct@rev.desc.d7.sspro.tau-T-by-type.p1.zoomclick-d3.icon.tau.tost5.p\endcsname{99.4}
\expandafter\def\csname ppt@rev.desc.d7.sspro.tau-T-by-type.p1.zoomclick-d3.icon.tau.tost5.p\endcsname{$+$99.4}
\expandafter\def\csname val@rev.desc.d7.sspro.tau-T-by-type.p1.zoomclick-d3.text.acc-F\endcsname{0.956}
\expandafter\def\csname cnt@rev.desc.d7.sspro.tau-T-by-type.p1.zoomclick-d3.text.acc-F\endcsname{1}
\expandafter\def\csname pct@rev.desc.d7.sspro.tau-T-by-type.p1.zoomclick-d3.text.acc-F\endcsname{95.6}
\expandafter\def\csname ppt@rev.desc.d7.sspro.tau-T-by-type.p1.zoomclick-d3.text.acc-F\endcsname{$+$95.6}
\expandafter\def\csname val@rev.desc.d7.sspro.tau-T-by-type.p1.zoomclick-d3.text.acc-direct\endcsname{0.779}
\expandafter\def\csname cnt@rev.desc.d7.sspro.tau-T-by-type.p1.zoomclick-d3.text.acc-direct\endcsname{1}
\expandafter\def\csname pct@rev.desc.d7.sspro.tau-T-by-type.p1.zoomclick-d3.text.acc-direct\endcsname{77.9}
\expandafter\def\csname ppt@rev.desc.d7.sspro.tau-T-by-type.p1.zoomclick-d3.text.acc-direct\endcsname{$+$77.9}
\expandafter\def\csname val@rev.desc.d7.sspro.tau-T-by-type.p1.zoomclick-d3.text.n-apps\endcsname{12}
\expandafter\def\csname cnt@rev.desc.d7.sspro.tau-T-by-type.p1.zoomclick-d3.text.n-apps\endcsname{12}
\expandafter\def\csname pct@rev.desc.d7.sspro.tau-T-by-type.p1.zoomclick-d3.text.n-apps\endcsname{1200.0}
\expandafter\def\csname ppt@rev.desc.d7.sspro.tau-T-by-type.p1.zoomclick-d3.text.n-apps\endcsname{$+$1200.0}
\expandafter\def\csname val@rev.desc.d7.sspro.tau-T-by-type.p1.zoomclick-d3.text.n-tasks\endcsname{68}
\expandafter\def\csname cnt@rev.desc.d7.sspro.tau-T-by-type.p1.zoomclick-d3.text.n-tasks\endcsname{68}
\expandafter\def\csname pct@rev.desc.d7.sspro.tau-T-by-type.p1.zoomclick-d3.text.n-tasks\endcsname{6800.0}
\expandafter\def\csname ppt@rev.desc.d7.sspro.tau-T-by-type.p1.zoomclick-d3.text.n-tasks\endcsname{$+$6800.0}
\expandafter\def\csname val@rev.desc.d7.sspro.tau-T-by-type.p1.zoomclick-d3.text.tau.ci90.0\endcsname{0.052}
\expandafter\def\csname cnt@rev.desc.d7.sspro.tau-T-by-type.p1.zoomclick-d3.text.tau.ci90.0\endcsname{0}
\expandafter\def\csname pct@rev.desc.d7.sspro.tau-T-by-type.p1.zoomclick-d3.text.tau.ci90.0\endcsname{5.2}
\expandafter\def\csname ppt@rev.desc.d7.sspro.tau-T-by-type.p1.zoomclick-d3.text.tau.ci90.0\endcsname{$+$5.2}
\expandafter\def\csname val@rev.desc.d7.sspro.tau-T-by-type.p1.zoomclick-d3.text.tau.ci90.1\endcsname{0.239}
\expandafter\def\csname cnt@rev.desc.d7.sspro.tau-T-by-type.p1.zoomclick-d3.text.tau.ci90.1\endcsname{0}
\expandafter\def\csname pct@rev.desc.d7.sspro.tau-T-by-type.p1.zoomclick-d3.text.tau.ci90.1\endcsname{23.9}
\expandafter\def\csname ppt@rev.desc.d7.sspro.tau-T-by-type.p1.zoomclick-d3.text.tau.ci90.1\endcsname{$+$23.9}
\expandafter\def\csname val@rev.desc.d7.sspro.tau-T-by-type.p1.zoomclick-d3.text.tau.ci95-excludes-0\endcsname{1}
\expandafter\def\csname cnt@rev.desc.d7.sspro.tau-T-by-type.p1.zoomclick-d3.text.tau.ci95-excludes-0\endcsname{1}
\expandafter\def\csname pct@rev.desc.d7.sspro.tau-T-by-type.p1.zoomclick-d3.text.tau.ci95-excludes-0\endcsname{100.0}
\expandafter\def\csname ppt@rev.desc.d7.sspro.tau-T-by-type.p1.zoomclick-d3.text.tau.ci95-excludes-0\endcsname{$+$100.0}
\expandafter\def\csname val@rev.desc.d7.sspro.tau-T-by-type.p1.zoomclick-d3.text.tau.ci95.0\endcsname{0.040}
\expandafter\def\csname cnt@rev.desc.d7.sspro.tau-T-by-type.p1.zoomclick-d3.text.tau.ci95.0\endcsname{0}
\expandafter\def\csname pct@rev.desc.d7.sspro.tau-T-by-type.p1.zoomclick-d3.text.tau.ci95.0\endcsname{4.0}
\expandafter\def\csname ppt@rev.desc.d7.sspro.tau-T-by-type.p1.zoomclick-d3.text.tau.ci95.0\endcsname{$+$4.0}
\expandafter\def\csname val@rev.desc.d7.sspro.tau-T-by-type.p1.zoomclick-d3.text.tau.ci95.1\endcsname{0.259}
\expandafter\def\csname cnt@rev.desc.d7.sspro.tau-T-by-type.p1.zoomclick-d3.text.tau.ci95.1\endcsname{0}
\expandafter\def\csname pct@rev.desc.d7.sspro.tau-T-by-type.p1.zoomclick-d3.text.tau.ci95.1\endcsname{25.9}
\expandafter\def\csname ppt@rev.desc.d7.sspro.tau-T-by-type.p1.zoomclick-d3.text.tau.ci95.1\endcsname{$+$25.9}
\expandafter\def\csname val@rev.desc.d7.sspro.tau-T-by-type.p1.zoomclick-d3.text.tau.mean\endcsname{0.138}
\expandafter\def\csname cnt@rev.desc.d7.sspro.tau-T-by-type.p1.zoomclick-d3.text.tau.mean\endcsname{0}
\expandafter\def\csname pct@rev.desc.d7.sspro.tau-T-by-type.p1.zoomclick-d3.text.tau.mean\endcsname{13.8}
\expandafter\def\csname ppt@rev.desc.d7.sspro.tau-T-by-type.p1.zoomclick-d3.text.tau.mean\endcsname{$+$13.8}
\expandafter\def\csname val@rev.desc.d7.sspro.tau-T-by-type.p1.zoomclick-d3.text.tau.n-apps\endcsname{12}
\expandafter\def\csname cnt@rev.desc.d7.sspro.tau-T-by-type.p1.zoomclick-d3.text.tau.n-apps\endcsname{12}
\expandafter\def\csname pct@rev.desc.d7.sspro.tau-T-by-type.p1.zoomclick-d3.text.tau.n-apps\endcsname{1200.0}
\expandafter\def\csname ppt@rev.desc.d7.sspro.tau-T-by-type.p1.zoomclick-d3.text.tau.n-apps\endcsname{$+$1200.0}
\expandafter\def\csname val@rev.desc.d7.sspro.tau-T-by-type.p1.zoomclick-d3.text.tau.tost1.ci90-inside\endcsname{0}
\expandafter\def\csname cnt@rev.desc.d7.sspro.tau-T-by-type.p1.zoomclick-d3.text.tau.tost1.ci90-inside\endcsname{0}
\expandafter\def\csname pct@rev.desc.d7.sspro.tau-T-by-type.p1.zoomclick-d3.text.tau.tost1.ci90-inside\endcsname{0.0}
\expandafter\def\csname ppt@rev.desc.d7.sspro.tau-T-by-type.p1.zoomclick-d3.text.tau.tost1.ci90-inside\endcsname{0.0}
\expandafter\def\csname val@rev.desc.d7.sspro.tau-T-by-type.p1.zoomclick-d3.text.tau.tost1.equivalent\endcsname{0}
\expandafter\def\csname cnt@rev.desc.d7.sspro.tau-T-by-type.p1.zoomclick-d3.text.tau.tost1.equivalent\endcsname{0}
\expandafter\def\csname pct@rev.desc.d7.sspro.tau-T-by-type.p1.zoomclick-d3.text.tau.tost1.equivalent\endcsname{0.0}
\expandafter\def\csname ppt@rev.desc.d7.sspro.tau-T-by-type.p1.zoomclick-d3.text.tau.tost1.equivalent\endcsname{0.0}
\expandafter\def\csname val@rev.desc.d7.sspro.tau-T-by-type.p1.zoomclick-d3.text.tau.tost1.p\endcsname{0.997}
\expandafter\def\csname cnt@rev.desc.d7.sspro.tau-T-by-type.p1.zoomclick-d3.text.tau.tost1.p\endcsname{1}
\expandafter\def\csname pct@rev.desc.d7.sspro.tau-T-by-type.p1.zoomclick-d3.text.tau.tost1.p\endcsname{99.7}
\expandafter\def\csname ppt@rev.desc.d7.sspro.tau-T-by-type.p1.zoomclick-d3.text.tau.tost1.p\endcsname{$+$99.7}
\expandafter\def\csname val@rev.desc.d7.sspro.tau-T-by-type.p1.zoomclick-d3.text.tau.tost5.ci90-inside\endcsname{0}
\expandafter\def\csname cnt@rev.desc.d7.sspro.tau-T-by-type.p1.zoomclick-d3.text.tau.tost5.ci90-inside\endcsname{0}
\expandafter\def\csname pct@rev.desc.d7.sspro.tau-T-by-type.p1.zoomclick-d3.text.tau.tost5.ci90-inside\endcsname{0.0}
\expandafter\def\csname ppt@rev.desc.d7.sspro.tau-T-by-type.p1.zoomclick-d3.text.tau.tost5.ci90-inside\endcsname{0.0}
\expandafter\def\csname val@rev.desc.d7.sspro.tau-T-by-type.p1.zoomclick-d3.text.tau.tost5.equivalent\endcsname{0}
\expandafter\def\csname cnt@rev.desc.d7.sspro.tau-T-by-type.p1.zoomclick-d3.text.tau.tost5.equivalent\endcsname{0}
\expandafter\def\csname pct@rev.desc.d7.sspro.tau-T-by-type.p1.zoomclick-d3.text.tau.tost5.equivalent\endcsname{0.0}
\expandafter\def\csname ppt@rev.desc.d7.sspro.tau-T-by-type.p1.zoomclick-d3.text.tau.tost5.equivalent\endcsname{0.0}
\expandafter\def\csname val@rev.desc.d7.sspro.tau-T-by-type.p1.zoomclick-d3.text.tau.tost5.p\endcsname{0.955}
\expandafter\def\csname cnt@rev.desc.d7.sspro.tau-T-by-type.p1.zoomclick-d3.text.tau.tost5.p\endcsname{1}
\expandafter\def\csname pct@rev.desc.d7.sspro.tau-T-by-type.p1.zoomclick-d3.text.tau.tost5.p\endcsname{95.5}
\expandafter\def\csname ppt@rev.desc.d7.sspro.tau-T-by-type.p1.zoomclick-d3.text.tau.tost5.p\endcsname{$+$95.5}
\expandafter\def\csname val@rev.desc.d7.sspro.tau-T-by-type.p2.center-d1.icon-minus-text-on-apps-with-both.ci90.0\endcsname{0.053}
\expandafter\def\csname cnt@rev.desc.d7.sspro.tau-T-by-type.p2.center-d1.icon-minus-text-on-apps-with-both.ci90.0\endcsname{0}
\expandafter\def\csname pct@rev.desc.d7.sspro.tau-T-by-type.p2.center-d1.icon-minus-text-on-apps-with-both.ci90.0\endcsname{5.3}
\expandafter\def\csname ppt@rev.desc.d7.sspro.tau-T-by-type.p2.center-d1.icon-minus-text-on-apps-with-both.ci90.0\endcsname{$+$5.3}
\expandafter\def\csname val@rev.desc.d7.sspro.tau-T-by-type.p2.center-d1.icon-minus-text-on-apps-with-both.ci90.1\endcsname{0.420}
\expandafter\def\csname cnt@rev.desc.d7.sspro.tau-T-by-type.p2.center-d1.icon-minus-text-on-apps-with-both.ci90.1\endcsname{0}
\expandafter\def\csname pct@rev.desc.d7.sspro.tau-T-by-type.p2.center-d1.icon-minus-text-on-apps-with-both.ci90.1\endcsname{42.0}
\expandafter\def\csname ppt@rev.desc.d7.sspro.tau-T-by-type.p2.center-d1.icon-minus-text-on-apps-with-both.ci90.1\endcsname{$+$42.0}
\expandafter\def\csname val@rev.desc.d7.sspro.tau-T-by-type.p2.center-d1.icon-minus-text-on-apps-with-both.ci95-excludes-0\endcsname{1}
\expandafter\def\csname cnt@rev.desc.d7.sspro.tau-T-by-type.p2.center-d1.icon-minus-text-on-apps-with-both.ci95-excludes-0\endcsname{1}
\expandafter\def\csname pct@rev.desc.d7.sspro.tau-T-by-type.p2.center-d1.icon-minus-text-on-apps-with-both.ci95-excludes-0\endcsname{100.0}
\expandafter\def\csname ppt@rev.desc.d7.sspro.tau-T-by-type.p2.center-d1.icon-minus-text-on-apps-with-both.ci95-excludes-0\endcsname{$+$100.0}
\expandafter\def\csname val@rev.desc.d7.sspro.tau-T-by-type.p2.center-d1.icon-minus-text-on-apps-with-both.ci95.0\endcsname{0.024}
\expandafter\def\csname cnt@rev.desc.d7.sspro.tau-T-by-type.p2.center-d1.icon-minus-text-on-apps-with-both.ci95.0\endcsname{0}
\expandafter\def\csname pct@rev.desc.d7.sspro.tau-T-by-type.p2.center-d1.icon-minus-text-on-apps-with-both.ci95.0\endcsname{2.4}
\expandafter\def\csname ppt@rev.desc.d7.sspro.tau-T-by-type.p2.center-d1.icon-minus-text-on-apps-with-both.ci95.0\endcsname{$+$2.4}
\expandafter\def\csname val@rev.desc.d7.sspro.tau-T-by-type.p2.center-d1.icon-minus-text-on-apps-with-both.ci95.1\endcsname{0.467}
\expandafter\def\csname cnt@rev.desc.d7.sspro.tau-T-by-type.p2.center-d1.icon-minus-text-on-apps-with-both.ci95.1\endcsname{0}
\expandafter\def\csname pct@rev.desc.d7.sspro.tau-T-by-type.p2.center-d1.icon-minus-text-on-apps-with-both.ci95.1\endcsname{46.7}
\expandafter\def\csname ppt@rev.desc.d7.sspro.tau-T-by-type.p2.center-d1.icon-minus-text-on-apps-with-both.ci95.1\endcsname{$+$46.7}
\expandafter\def\csname val@rev.desc.d7.sspro.tau-T-by-type.p2.center-d1.icon-minus-text-on-apps-with-both.mean\endcsname{0.222}
\expandafter\def\csname cnt@rev.desc.d7.sspro.tau-T-by-type.p2.center-d1.icon-minus-text-on-apps-with-both.mean\endcsname{0}
\expandafter\def\csname pct@rev.desc.d7.sspro.tau-T-by-type.p2.center-d1.icon-minus-text-on-apps-with-both.mean\endcsname{22.2}
\expandafter\def\csname ppt@rev.desc.d7.sspro.tau-T-by-type.p2.center-d1.icon-minus-text-on-apps-with-both.mean\endcsname{$+$22.2}
\expandafter\def\csname val@rev.desc.d7.sspro.tau-T-by-type.p2.center-d1.icon-minus-text-on-apps-with-both.n-apps\endcsname{10}
\expandafter\def\csname cnt@rev.desc.d7.sspro.tau-T-by-type.p2.center-d1.icon-minus-text-on-apps-with-both.n-apps\endcsname{10}
\expandafter\def\csname pct@rev.desc.d7.sspro.tau-T-by-type.p2.center-d1.icon-minus-text-on-apps-with-both.n-apps\endcsname{1000.0}
\expandafter\def\csname ppt@rev.desc.d7.sspro.tau-T-by-type.p2.center-d1.icon-minus-text-on-apps-with-both.n-apps\endcsname{$+$1000.0}
\expandafter\def\csname val@rev.desc.d7.sspro.tau-T-by-type.p2.center-d1.icon-minus-text-on-apps-with-both.tost1.ci90-inside\endcsname{0}
\expandafter\def\csname cnt@rev.desc.d7.sspro.tau-T-by-type.p2.center-d1.icon-minus-text-on-apps-with-both.tost1.ci90-inside\endcsname{0}
\expandafter\def\csname pct@rev.desc.d7.sspro.tau-T-by-type.p2.center-d1.icon-minus-text-on-apps-with-both.tost1.ci90-inside\endcsname{0.0}
\expandafter\def\csname ppt@rev.desc.d7.sspro.tau-T-by-type.p2.center-d1.icon-minus-text-on-apps-with-both.tost1.ci90-inside\endcsname{0.0}
\expandafter\def\csname val@rev.desc.d7.sspro.tau-T-by-type.p2.center-d1.icon-minus-text-on-apps-with-both.tost1.equivalent\endcsname{0}
\expandafter\def\csname cnt@rev.desc.d7.sspro.tau-T-by-type.p2.center-d1.icon-minus-text-on-apps-with-both.tost1.equivalent\endcsname{0}
\expandafter\def\csname pct@rev.desc.d7.sspro.tau-T-by-type.p2.center-d1.icon-minus-text-on-apps-with-both.tost1.equivalent\endcsname{0.0}
\expandafter\def\csname ppt@rev.desc.d7.sspro.tau-T-by-type.p2.center-d1.icon-minus-text-on-apps-with-both.tost1.equivalent\endcsname{0.0}
\expandafter\def\csname val@rev.desc.d7.sspro.tau-T-by-type.p2.center-d1.icon-minus-text-on-apps-with-both.tost1.p\endcsname{0.983}
\expandafter\def\csname cnt@rev.desc.d7.sspro.tau-T-by-type.p2.center-d1.icon-minus-text-on-apps-with-both.tost1.p\endcsname{1}
\expandafter\def\csname pct@rev.desc.d7.sspro.tau-T-by-type.p2.center-d1.icon-minus-text-on-apps-with-both.tost1.p\endcsname{98.3}
\expandafter\def\csname ppt@rev.desc.d7.sspro.tau-T-by-type.p2.center-d1.icon-minus-text-on-apps-with-both.tost1.p\endcsname{$+$98.3}
\expandafter\def\csname val@rev.desc.d7.sspro.tau-T-by-type.p2.center-d1.icon-minus-text-on-apps-with-both.tost5.ci90-inside\endcsname{0}
\expandafter\def\csname cnt@rev.desc.d7.sspro.tau-T-by-type.p2.center-d1.icon-minus-text-on-apps-with-both.tost5.ci90-inside\endcsname{0}
\expandafter\def\csname pct@rev.desc.d7.sspro.tau-T-by-type.p2.center-d1.icon-minus-text-on-apps-with-both.tost5.ci90-inside\endcsname{0.0}
\expandafter\def\csname ppt@rev.desc.d7.sspro.tau-T-by-type.p2.center-d1.icon-minus-text-on-apps-with-both.tost5.ci90-inside\endcsname{0.0}
\expandafter\def\csname val@rev.desc.d7.sspro.tau-T-by-type.p2.center-d1.icon-minus-text-on-apps-with-both.tost5.equivalent\endcsname{0}
\expandafter\def\csname cnt@rev.desc.d7.sspro.tau-T-by-type.p2.center-d1.icon-minus-text-on-apps-with-both.tost5.equivalent\endcsname{0}
\expandafter\def\csname pct@rev.desc.d7.sspro.tau-T-by-type.p2.center-d1.icon-minus-text-on-apps-with-both.tost5.equivalent\endcsname{0.0}
\expandafter\def\csname ppt@rev.desc.d7.sspro.tau-T-by-type.p2.center-d1.icon-minus-text-on-apps-with-both.tost5.equivalent\endcsname{0.0}
\expandafter\def\csname val@rev.desc.d7.sspro.tau-T-by-type.p2.center-d1.icon-minus-text-on-apps-with-both.tost5.p\endcsname{0.954}
\expandafter\def\csname cnt@rev.desc.d7.sspro.tau-T-by-type.p2.center-d1.icon-minus-text-on-apps-with-both.tost5.p\endcsname{1}
\expandafter\def\csname pct@rev.desc.d7.sspro.tau-T-by-type.p2.center-d1.icon-minus-text-on-apps-with-both.tost5.p\endcsname{95.4}
\expandafter\def\csname ppt@rev.desc.d7.sspro.tau-T-by-type.p2.center-d1.icon-minus-text-on-apps-with-both.tost5.p\endcsname{$+$95.4}
\expandafter\def\csname val@rev.desc.d7.sspro.tau-T-by-type.p2.center-d1.icon.acc-F\endcsname{0.786}
\expandafter\def\csname cnt@rev.desc.d7.sspro.tau-T-by-type.p2.center-d1.icon.acc-F\endcsname{1}
\expandafter\def\csname pct@rev.desc.d7.sspro.tau-T-by-type.p2.center-d1.icon.acc-F\endcsname{78.6}
\expandafter\def\csname ppt@rev.desc.d7.sspro.tau-T-by-type.p2.center-d1.icon.acc-F\endcsname{$+$78.6}
\expandafter\def\csname val@rev.desc.d7.sspro.tau-T-by-type.p2.center-d1.icon.acc-direct\endcsname{0.571}
\expandafter\def\csname cnt@rev.desc.d7.sspro.tau-T-by-type.p2.center-d1.icon.acc-direct\endcsname{1}
\expandafter\def\csname pct@rev.desc.d7.sspro.tau-T-by-type.p2.center-d1.icon.acc-direct\endcsname{57.1}
\expandafter\def\csname ppt@rev.desc.d7.sspro.tau-T-by-type.p2.center-d1.icon.acc-direct\endcsname{$+$57.1}
\expandafter\def\csname val@rev.desc.d7.sspro.tau-T-by-type.p2.center-d1.icon.n-apps\endcsname{10}
\expandafter\def\csname cnt@rev.desc.d7.sspro.tau-T-by-type.p2.center-d1.icon.n-apps\endcsname{10}
\expandafter\def\csname pct@rev.desc.d7.sspro.tau-T-by-type.p2.center-d1.icon.n-apps\endcsname{1000.0}
\expandafter\def\csname ppt@rev.desc.d7.sspro.tau-T-by-type.p2.center-d1.icon.n-apps\endcsname{$+$1000.0}
\expandafter\def\csname val@rev.desc.d7.sspro.tau-T-by-type.p2.center-d1.icon.n-tasks\endcsname{28}
\expandafter\def\csname cnt@rev.desc.d7.sspro.tau-T-by-type.p2.center-d1.icon.n-tasks\endcsname{28}
\expandafter\def\csname pct@rev.desc.d7.sspro.tau-T-by-type.p2.center-d1.icon.n-tasks\endcsname{2800.0}
\expandafter\def\csname ppt@rev.desc.d7.sspro.tau-T-by-type.p2.center-d1.icon.n-tasks\endcsname{$+$2800.0}
\expandafter\def\csname val@rev.desc.d7.sspro.tau-T-by-type.p2.center-d1.icon.tau.ci90.0\endcsname{0.113}
\expandafter\def\csname cnt@rev.desc.d7.sspro.tau-T-by-type.p2.center-d1.icon.tau.ci90.0\endcsname{0}
\expandafter\def\csname pct@rev.desc.d7.sspro.tau-T-by-type.p2.center-d1.icon.tau.ci90.0\endcsname{11.3}
\expandafter\def\csname ppt@rev.desc.d7.sspro.tau-T-by-type.p2.center-d1.icon.tau.ci90.0\endcsname{$+$11.3}
\expandafter\def\csname val@rev.desc.d7.sspro.tau-T-by-type.p2.center-d1.icon.tau.ci90.1\endcsname{0.453}
\expandafter\def\csname cnt@rev.desc.d7.sspro.tau-T-by-type.p2.center-d1.icon.tau.ci90.1\endcsname{0}
\expandafter\def\csname pct@rev.desc.d7.sspro.tau-T-by-type.p2.center-d1.icon.tau.ci90.1\endcsname{45.3}
\expandafter\def\csname ppt@rev.desc.d7.sspro.tau-T-by-type.p2.center-d1.icon.tau.ci90.1\endcsname{$+$45.3}
\expandafter\def\csname val@rev.desc.d7.sspro.tau-T-by-type.p2.center-d1.icon.tau.ci95-excludes-0\endcsname{1}
\expandafter\def\csname cnt@rev.desc.d7.sspro.tau-T-by-type.p2.center-d1.icon.tau.ci95-excludes-0\endcsname{1}
\expandafter\def\csname pct@rev.desc.d7.sspro.tau-T-by-type.p2.center-d1.icon.tau.ci95-excludes-0\endcsname{100.0}
\expandafter\def\csname ppt@rev.desc.d7.sspro.tau-T-by-type.p2.center-d1.icon.tau.ci95-excludes-0\endcsname{$+$100.0}
\expandafter\def\csname val@rev.desc.d7.sspro.tau-T-by-type.p2.center-d1.icon.tau.ci95.0\endcsname{0.087}
\expandafter\def\csname cnt@rev.desc.d7.sspro.tau-T-by-type.p2.center-d1.icon.tau.ci95.0\endcsname{0}
\expandafter\def\csname pct@rev.desc.d7.sspro.tau-T-by-type.p2.center-d1.icon.tau.ci95.0\endcsname{8.7}
\expandafter\def\csname ppt@rev.desc.d7.sspro.tau-T-by-type.p2.center-d1.icon.tau.ci95.0\endcsname{$+$8.7}
\expandafter\def\csname val@rev.desc.d7.sspro.tau-T-by-type.p2.center-d1.icon.tau.ci95.1\endcsname{0.487}
\expandafter\def\csname cnt@rev.desc.d7.sspro.tau-T-by-type.p2.center-d1.icon.tau.ci95.1\endcsname{0}
\expandafter\def\csname pct@rev.desc.d7.sspro.tau-T-by-type.p2.center-d1.icon.tau.ci95.1\endcsname{48.7}
\expandafter\def\csname ppt@rev.desc.d7.sspro.tau-T-by-type.p2.center-d1.icon.tau.ci95.1\endcsname{$+$48.7}
\expandafter\def\csname val@rev.desc.d7.sspro.tau-T-by-type.p2.center-d1.icon.tau.mean\endcsname{0.270}
\expandafter\def\csname cnt@rev.desc.d7.sspro.tau-T-by-type.p2.center-d1.icon.tau.mean\endcsname{0}
\expandafter\def\csname pct@rev.desc.d7.sspro.tau-T-by-type.p2.center-d1.icon.tau.mean\endcsname{27.0}
\expandafter\def\csname ppt@rev.desc.d7.sspro.tau-T-by-type.p2.center-d1.icon.tau.mean\endcsname{$+$27.0}
\expandafter\def\csname val@rev.desc.d7.sspro.tau-T-by-type.p2.center-d1.icon.tau.n-apps\endcsname{10}
\expandafter\def\csname cnt@rev.desc.d7.sspro.tau-T-by-type.p2.center-d1.icon.tau.n-apps\endcsname{10}
\expandafter\def\csname pct@rev.desc.d7.sspro.tau-T-by-type.p2.center-d1.icon.tau.n-apps\endcsname{1000.0}
\expandafter\def\csname ppt@rev.desc.d7.sspro.tau-T-by-type.p2.center-d1.icon.tau.n-apps\endcsname{$+$1000.0}
\expandafter\def\csname val@rev.desc.d7.sspro.tau-T-by-type.p2.center-d1.icon.tau.tost1.ci90-inside\endcsname{0}
\expandafter\def\csname cnt@rev.desc.d7.sspro.tau-T-by-type.p2.center-d1.icon.tau.tost1.ci90-inside\endcsname{0}
\expandafter\def\csname pct@rev.desc.d7.sspro.tau-T-by-type.p2.center-d1.icon.tau.tost1.ci90-inside\endcsname{0.0}
\expandafter\def\csname ppt@rev.desc.d7.sspro.tau-T-by-type.p2.center-d1.icon.tau.tost1.ci90-inside\endcsname{0.0}
\expandafter\def\csname val@rev.desc.d7.sspro.tau-T-by-type.p2.center-d1.icon.tau.tost1.equivalent\endcsname{0}
\expandafter\def\csname cnt@rev.desc.d7.sspro.tau-T-by-type.p2.center-d1.icon.tau.tost1.equivalent\endcsname{0}
\expandafter\def\csname pct@rev.desc.d7.sspro.tau-T-by-type.p2.center-d1.icon.tau.tost1.equivalent\endcsname{0.0}
\expandafter\def\csname ppt@rev.desc.d7.sspro.tau-T-by-type.p2.center-d1.icon.tau.tost1.equivalent\endcsname{0.0}
\expandafter\def\csname val@rev.desc.d7.sspro.tau-T-by-type.p2.center-d1.icon.tau.tost1.p\endcsname{0.999}
\expandafter\def\csname cnt@rev.desc.d7.sspro.tau-T-by-type.p2.center-d1.icon.tau.tost1.p\endcsname{1}
\expandafter\def\csname pct@rev.desc.d7.sspro.tau-T-by-type.p2.center-d1.icon.tau.tost1.p\endcsname{99.9}
\expandafter\def\csname ppt@rev.desc.d7.sspro.tau-T-by-type.p2.center-d1.icon.tau.tost1.p\endcsname{$+$99.9}
\expandafter\def\csname val@rev.desc.d7.sspro.tau-T-by-type.p2.center-d1.icon.tau.tost5.ci90-inside\endcsname{0}
\expandafter\def\csname cnt@rev.desc.d7.sspro.tau-T-by-type.p2.center-d1.icon.tau.tost5.ci90-inside\endcsname{0}
\expandafter\def\csname pct@rev.desc.d7.sspro.tau-T-by-type.p2.center-d1.icon.tau.tost5.ci90-inside\endcsname{0.0}
\expandafter\def\csname ppt@rev.desc.d7.sspro.tau-T-by-type.p2.center-d1.icon.tau.tost5.ci90-inside\endcsname{0.0}
\expandafter\def\csname val@rev.desc.d7.sspro.tau-T-by-type.p2.center-d1.icon.tau.tost5.equivalent\endcsname{0}
\expandafter\def\csname cnt@rev.desc.d7.sspro.tau-T-by-type.p2.center-d1.icon.tau.tost5.equivalent\endcsname{0}
\expandafter\def\csname pct@rev.desc.d7.sspro.tau-T-by-type.p2.center-d1.icon.tau.tost5.equivalent\endcsname{0.0}
\expandafter\def\csname ppt@rev.desc.d7.sspro.tau-T-by-type.p2.center-d1.icon.tau.tost5.equivalent\endcsname{0.0}
\expandafter\def\csname val@rev.desc.d7.sspro.tau-T-by-type.p2.center-d1.icon.tau.tost5.p\endcsname{0.994}
\expandafter\def\csname cnt@rev.desc.d7.sspro.tau-T-by-type.p2.center-d1.icon.tau.tost5.p\endcsname{1}
\expandafter\def\csname pct@rev.desc.d7.sspro.tau-T-by-type.p2.center-d1.icon.tau.tost5.p\endcsname{99.4}
\expandafter\def\csname ppt@rev.desc.d7.sspro.tau-T-by-type.p2.center-d1.icon.tau.tost5.p\endcsname{$+$99.4}
\expandafter\def\csname val@rev.desc.d7.sspro.tau-T-by-type.p2.center-d1.text.acc-F\endcsname{0.971}
\expandafter\def\csname cnt@rev.desc.d7.sspro.tau-T-by-type.p2.center-d1.text.acc-F\endcsname{1}
\expandafter\def\csname pct@rev.desc.d7.sspro.tau-T-by-type.p2.center-d1.text.acc-F\endcsname{97.1}
\expandafter\def\csname ppt@rev.desc.d7.sspro.tau-T-by-type.p2.center-d1.text.acc-F\endcsname{$+$97.1}
\expandafter\def\csname val@rev.desc.d7.sspro.tau-T-by-type.p2.center-d1.text.acc-direct\endcsname{0.838}
\expandafter\def\csname cnt@rev.desc.d7.sspro.tau-T-by-type.p2.center-d1.text.acc-direct\endcsname{1}
\expandafter\def\csname pct@rev.desc.d7.sspro.tau-T-by-type.p2.center-d1.text.acc-direct\endcsname{83.8}
\expandafter\def\csname ppt@rev.desc.d7.sspro.tau-T-by-type.p2.center-d1.text.acc-direct\endcsname{$+$83.8}
\expandafter\def\csname val@rev.desc.d7.sspro.tau-T-by-type.p2.center-d1.text.n-apps\endcsname{12}
\expandafter\def\csname cnt@rev.desc.d7.sspro.tau-T-by-type.p2.center-d1.text.n-apps\endcsname{12}
\expandafter\def\csname pct@rev.desc.d7.sspro.tau-T-by-type.p2.center-d1.text.n-apps\endcsname{1200.0}
\expandafter\def\csname ppt@rev.desc.d7.sspro.tau-T-by-type.p2.center-d1.text.n-apps\endcsname{$+$1200.0}
\expandafter\def\csname val@rev.desc.d7.sspro.tau-T-by-type.p2.center-d1.text.n-tasks\endcsname{68}
\expandafter\def\csname cnt@rev.desc.d7.sspro.tau-T-by-type.p2.center-d1.text.n-tasks\endcsname{68}
\expandafter\def\csname pct@rev.desc.d7.sspro.tau-T-by-type.p2.center-d1.text.n-tasks\endcsname{6800.0}
\expandafter\def\csname ppt@rev.desc.d7.sspro.tau-T-by-type.p2.center-d1.text.n-tasks\endcsname{$+$6800.0}
\expandafter\def\csname val@rev.desc.d7.sspro.tau-T-by-type.p2.center-d1.text.tau.ci90.0\endcsname{0.040}
\expandafter\def\csname cnt@rev.desc.d7.sspro.tau-T-by-type.p2.center-d1.text.tau.ci90.0\endcsname{0}
\expandafter\def\csname pct@rev.desc.d7.sspro.tau-T-by-type.p2.center-d1.text.tau.ci90.0\endcsname{4.0}
\expandafter\def\csname ppt@rev.desc.d7.sspro.tau-T-by-type.p2.center-d1.text.tau.ci90.0\endcsname{$+$4.0}
\expandafter\def\csname val@rev.desc.d7.sspro.tau-T-by-type.p2.center-d1.text.tau.ci90.1\endcsname{0.180}
\expandafter\def\csname cnt@rev.desc.d7.sspro.tau-T-by-type.p2.center-d1.text.tau.ci90.1\endcsname{0}
\expandafter\def\csname pct@rev.desc.d7.sspro.tau-T-by-type.p2.center-d1.text.tau.ci90.1\endcsname{18.0}
\expandafter\def\csname ppt@rev.desc.d7.sspro.tau-T-by-type.p2.center-d1.text.tau.ci90.1\endcsname{$+$18.0}
\expandafter\def\csname val@rev.desc.d7.sspro.tau-T-by-type.p2.center-d1.text.tau.ci95-excludes-0\endcsname{1}
\expandafter\def\csname cnt@rev.desc.d7.sspro.tau-T-by-type.p2.center-d1.text.tau.ci95-excludes-0\endcsname{1}
\expandafter\def\csname pct@rev.desc.d7.sspro.tau-T-by-type.p2.center-d1.text.tau.ci95-excludes-0\endcsname{100.0}
\expandafter\def\csname ppt@rev.desc.d7.sspro.tau-T-by-type.p2.center-d1.text.tau.ci95-excludes-0\endcsname{$+$100.0}
\expandafter\def\csname val@rev.desc.d7.sspro.tau-T-by-type.p2.center-d1.text.tau.ci95.0\endcsname{0.028}
\expandafter\def\csname cnt@rev.desc.d7.sspro.tau-T-by-type.p2.center-d1.text.tau.ci95.0\endcsname{0}
\expandafter\def\csname pct@rev.desc.d7.sspro.tau-T-by-type.p2.center-d1.text.tau.ci95.0\endcsname{2.8}
\expandafter\def\csname ppt@rev.desc.d7.sspro.tau-T-by-type.p2.center-d1.text.tau.ci95.0\endcsname{$+$2.8}
\expandafter\def\csname val@rev.desc.d7.sspro.tau-T-by-type.p2.center-d1.text.tau.ci95.1\endcsname{0.194}
\expandafter\def\csname cnt@rev.desc.d7.sspro.tau-T-by-type.p2.center-d1.text.tau.ci95.1\endcsname{0}
\expandafter\def\csname pct@rev.desc.d7.sspro.tau-T-by-type.p2.center-d1.text.tau.ci95.1\endcsname{19.4}
\expandafter\def\csname ppt@rev.desc.d7.sspro.tau-T-by-type.p2.center-d1.text.tau.ci95.1\endcsname{$+$19.4}
\expandafter\def\csname val@rev.desc.d7.sspro.tau-T-by-type.p2.center-d1.text.tau.mean\endcsname{0.102}
\expandafter\def\csname cnt@rev.desc.d7.sspro.tau-T-by-type.p2.center-d1.text.tau.mean\endcsname{0}
\expandafter\def\csname pct@rev.desc.d7.sspro.tau-T-by-type.p2.center-d1.text.tau.mean\endcsname{10.2}
\expandafter\def\csname ppt@rev.desc.d7.sspro.tau-T-by-type.p2.center-d1.text.tau.mean\endcsname{$+$10.2}
\expandafter\def\csname val@rev.desc.d7.sspro.tau-T-by-type.p2.center-d1.text.tau.n-apps\endcsname{12}
\expandafter\def\csname cnt@rev.desc.d7.sspro.tau-T-by-type.p2.center-d1.text.tau.n-apps\endcsname{12}
\expandafter\def\csname pct@rev.desc.d7.sspro.tau-T-by-type.p2.center-d1.text.tau.n-apps\endcsname{1200.0}
\expandafter\def\csname ppt@rev.desc.d7.sspro.tau-T-by-type.p2.center-d1.text.tau.n-apps\endcsname{$+$1200.0}
\expandafter\def\csname val@rev.desc.d7.sspro.tau-T-by-type.p2.center-d1.text.tau.tost1.ci90-inside\endcsname{0}
\expandafter\def\csname cnt@rev.desc.d7.sspro.tau-T-by-type.p2.center-d1.text.tau.tost1.ci90-inside\endcsname{0}
\expandafter\def\csname pct@rev.desc.d7.sspro.tau-T-by-type.p2.center-d1.text.tau.tost1.ci90-inside\endcsname{0.0}
\expandafter\def\csname ppt@rev.desc.d7.sspro.tau-T-by-type.p2.center-d1.text.tau.tost1.ci90-inside\endcsname{0.0}
\expandafter\def\csname val@rev.desc.d7.sspro.tau-T-by-type.p2.center-d1.text.tau.tost1.equivalent\endcsname{0}
\expandafter\def\csname cnt@rev.desc.d7.sspro.tau-T-by-type.p2.center-d1.text.tau.tost1.equivalent\endcsname{0}
\expandafter\def\csname pct@rev.desc.d7.sspro.tau-T-by-type.p2.center-d1.text.tau.tost1.equivalent\endcsname{0.0}
\expandafter\def\csname ppt@rev.desc.d7.sspro.tau-T-by-type.p2.center-d1.text.tau.tost1.equivalent\endcsname{0.0}
\expandafter\def\csname val@rev.desc.d7.sspro.tau-T-by-type.p2.center-d1.text.tau.tost1.p\endcsname{0.998}
\expandafter\def\csname cnt@rev.desc.d7.sspro.tau-T-by-type.p2.center-d1.text.tau.tost1.p\endcsname{1}
\expandafter\def\csname pct@rev.desc.d7.sspro.tau-T-by-type.p2.center-d1.text.tau.tost1.p\endcsname{99.8}
\expandafter\def\csname ppt@rev.desc.d7.sspro.tau-T-by-type.p2.center-d1.text.tau.tost1.p\endcsname{$+$99.8}
\expandafter\def\csname val@rev.desc.d7.sspro.tau-T-by-type.p2.center-d1.text.tau.tost5.ci90-inside\endcsname{0}
\expandafter\def\csname cnt@rev.desc.d7.sspro.tau-T-by-type.p2.center-d1.text.tau.tost5.ci90-inside\endcsname{0}
\expandafter\def\csname pct@rev.desc.d7.sspro.tau-T-by-type.p2.center-d1.text.tau.tost5.ci90-inside\endcsname{0.0}
\expandafter\def\csname ppt@rev.desc.d7.sspro.tau-T-by-type.p2.center-d1.text.tau.tost5.ci90-inside\endcsname{0.0}
\expandafter\def\csname val@rev.desc.d7.sspro.tau-T-by-type.p2.center-d1.text.tau.tost5.equivalent\endcsname{0}
\expandafter\def\csname cnt@rev.desc.d7.sspro.tau-T-by-type.p2.center-d1.text.tau.tost5.equivalent\endcsname{0}
\expandafter\def\csname pct@rev.desc.d7.sspro.tau-T-by-type.p2.center-d1.text.tau.tost5.equivalent\endcsname{0.0}
\expandafter\def\csname ppt@rev.desc.d7.sspro.tau-T-by-type.p2.center-d1.text.tau.tost5.equivalent\endcsname{0.0}
\expandafter\def\csname val@rev.desc.d7.sspro.tau-T-by-type.p2.center-d1.text.tau.tost5.p\endcsname{0.901}
\expandafter\def\csname cnt@rev.desc.d7.sspro.tau-T-by-type.p2.center-d1.text.tau.tost5.p\endcsname{1}
\expandafter\def\csname pct@rev.desc.d7.sspro.tau-T-by-type.p2.center-d1.text.tau.tost5.p\endcsname{90.1}
\expandafter\def\csname ppt@rev.desc.d7.sspro.tau-T-by-type.p2.center-d1.text.tau.tost5.p\endcsname{$+$90.1}
\expandafter\def\csname val@rev.desc.d7.sspro.tau-T-by-type.p2.context-d1.icon-minus-text-on-apps-with-both.ci90.0\endcsname{-0.067}
\expandafter\def\csname cnt@rev.desc.d7.sspro.tau-T-by-type.p2.context-d1.icon-minus-text-on-apps-with-both.ci90.0\endcsname{0}
\expandafter\def\csname pct@rev.desc.d7.sspro.tau-T-by-type.p2.context-d1.icon-minus-text-on-apps-with-both.ci90.0\endcsname{$-$6.7}
\expandafter\def\csname ppt@rev.desc.d7.sspro.tau-T-by-type.p2.context-d1.icon-minus-text-on-apps-with-both.ci90.0\endcsname{$-$6.7}
\expandafter\def\csname val@rev.desc.d7.sspro.tau-T-by-type.p2.context-d1.icon-minus-text-on-apps-with-both.ci90.1\endcsname{0.353}
\expandafter\def\csname cnt@rev.desc.d7.sspro.tau-T-by-type.p2.context-d1.icon-minus-text-on-apps-with-both.ci90.1\endcsname{0}
\expandafter\def\csname pct@rev.desc.d7.sspro.tau-T-by-type.p2.context-d1.icon-minus-text-on-apps-with-both.ci90.1\endcsname{35.3}
\expandafter\def\csname ppt@rev.desc.d7.sspro.tau-T-by-type.p2.context-d1.icon-minus-text-on-apps-with-both.ci90.1\endcsname{$+$35.3}
\expandafter\def\csname val@rev.desc.d7.sspro.tau-T-by-type.p2.context-d1.icon-minus-text-on-apps-with-both.ci95-excludes-0\endcsname{0}
\expandafter\def\csname cnt@rev.desc.d7.sspro.tau-T-by-type.p2.context-d1.icon-minus-text-on-apps-with-both.ci95-excludes-0\endcsname{0}
\expandafter\def\csname pct@rev.desc.d7.sspro.tau-T-by-type.p2.context-d1.icon-minus-text-on-apps-with-both.ci95-excludes-0\endcsname{0.0}
\expandafter\def\csname ppt@rev.desc.d7.sspro.tau-T-by-type.p2.context-d1.icon-minus-text-on-apps-with-both.ci95-excludes-0\endcsname{0.0}
\expandafter\def\csname val@rev.desc.d7.sspro.tau-T-by-type.p2.context-d1.icon-minus-text-on-apps-with-both.ci95.0\endcsname{-0.109}
\expandafter\def\csname cnt@rev.desc.d7.sspro.tau-T-by-type.p2.context-d1.icon-minus-text-on-apps-with-both.ci95.0\endcsname{0}
\expandafter\def\csname pct@rev.desc.d7.sspro.tau-T-by-type.p2.context-d1.icon-minus-text-on-apps-with-both.ci95.0\endcsname{$-$10.9}
\expandafter\def\csname ppt@rev.desc.d7.sspro.tau-T-by-type.p2.context-d1.icon-minus-text-on-apps-with-both.ci95.0\endcsname{$-$10.9}
\expandafter\def\csname val@rev.desc.d7.sspro.tau-T-by-type.p2.context-d1.icon-minus-text-on-apps-with-both.ci95.1\endcsname{0.400}
\expandafter\def\csname cnt@rev.desc.d7.sspro.tau-T-by-type.p2.context-d1.icon-minus-text-on-apps-with-both.ci95.1\endcsname{0}
\expandafter\def\csname pct@rev.desc.d7.sspro.tau-T-by-type.p2.context-d1.icon-minus-text-on-apps-with-both.ci95.1\endcsname{40.0}
\expandafter\def\csname ppt@rev.desc.d7.sspro.tau-T-by-type.p2.context-d1.icon-minus-text-on-apps-with-both.ci95.1\endcsname{$+$40.0}
\expandafter\def\csname val@rev.desc.d7.sspro.tau-T-by-type.p2.context-d1.icon-minus-text-on-apps-with-both.mean\endcsname{0.139}
\expandafter\def\csname cnt@rev.desc.d7.sspro.tau-T-by-type.p2.context-d1.icon-minus-text-on-apps-with-both.mean\endcsname{0}
\expandafter\def\csname pct@rev.desc.d7.sspro.tau-T-by-type.p2.context-d1.icon-minus-text-on-apps-with-both.mean\endcsname{13.9}
\expandafter\def\csname ppt@rev.desc.d7.sspro.tau-T-by-type.p2.context-d1.icon-minus-text-on-apps-with-both.mean\endcsname{$+$13.9}
\expandafter\def\csname val@rev.desc.d7.sspro.tau-T-by-type.p2.context-d1.icon-minus-text-on-apps-with-both.n-apps\endcsname{10}
\expandafter\def\csname cnt@rev.desc.d7.sspro.tau-T-by-type.p2.context-d1.icon-minus-text-on-apps-with-both.n-apps\endcsname{10}
\expandafter\def\csname pct@rev.desc.d7.sspro.tau-T-by-type.p2.context-d1.icon-minus-text-on-apps-with-both.n-apps\endcsname{1000.0}
\expandafter\def\csname ppt@rev.desc.d7.sspro.tau-T-by-type.p2.context-d1.icon-minus-text-on-apps-with-both.n-apps\endcsname{$+$1000.0}
\expandafter\def\csname val@rev.desc.d7.sspro.tau-T-by-type.p2.context-d1.icon-minus-text-on-apps-with-both.tost1.ci90-inside\endcsname{0}
\expandafter\def\csname cnt@rev.desc.d7.sspro.tau-T-by-type.p2.context-d1.icon-minus-text-on-apps-with-both.tost1.ci90-inside\endcsname{0}
\expandafter\def\csname pct@rev.desc.d7.sspro.tau-T-by-type.p2.context-d1.icon-minus-text-on-apps-with-both.tost1.ci90-inside\endcsname{0.0}
\expandafter\def\csname ppt@rev.desc.d7.sspro.tau-T-by-type.p2.context-d1.icon-minus-text-on-apps-with-both.tost1.ci90-inside\endcsname{0.0}
\expandafter\def\csname val@rev.desc.d7.sspro.tau-T-by-type.p2.context-d1.icon-minus-text-on-apps-with-both.tost1.equivalent\endcsname{0}
\expandafter\def\csname cnt@rev.desc.d7.sspro.tau-T-by-type.p2.context-d1.icon-minus-text-on-apps-with-both.tost1.equivalent\endcsname{0}
\expandafter\def\csname pct@rev.desc.d7.sspro.tau-T-by-type.p2.context-d1.icon-minus-text-on-apps-with-both.tost1.equivalent\endcsname{0.0}
\expandafter\def\csname ppt@rev.desc.d7.sspro.tau-T-by-type.p2.context-d1.icon-minus-text-on-apps-with-both.tost1.equivalent\endcsname{0.0}
\expandafter\def\csname val@rev.desc.d7.sspro.tau-T-by-type.p2.context-d1.icon-minus-text-on-apps-with-both.tost1.p\endcsname{0.852}
\expandafter\def\csname cnt@rev.desc.d7.sspro.tau-T-by-type.p2.context-d1.icon-minus-text-on-apps-with-both.tost1.p\endcsname{1}
\expandafter\def\csname pct@rev.desc.d7.sspro.tau-T-by-type.p2.context-d1.icon-minus-text-on-apps-with-both.tost1.p\endcsname{85.2}
\expandafter\def\csname ppt@rev.desc.d7.sspro.tau-T-by-type.p2.context-d1.icon-minus-text-on-apps-with-both.tost1.p\endcsname{$+$85.2}
\expandafter\def\csname val@rev.desc.d7.sspro.tau-T-by-type.p2.context-d1.icon-minus-text-on-apps-with-both.tost5.ci90-inside\endcsname{0}
\expandafter\def\csname cnt@rev.desc.d7.sspro.tau-T-by-type.p2.context-d1.icon-minus-text-on-apps-with-both.tost5.ci90-inside\endcsname{0}
\expandafter\def\csname pct@rev.desc.d7.sspro.tau-T-by-type.p2.context-d1.icon-minus-text-on-apps-with-both.tost5.ci90-inside\endcsname{0.0}
\expandafter\def\csname ppt@rev.desc.d7.sspro.tau-T-by-type.p2.context-d1.icon-minus-text-on-apps-with-both.tost5.ci90-inside\endcsname{0.0}
\expandafter\def\csname val@rev.desc.d7.sspro.tau-T-by-type.p2.context-d1.icon-minus-text-on-apps-with-both.tost5.equivalent\endcsname{0}
\expandafter\def\csname cnt@rev.desc.d7.sspro.tau-T-by-type.p2.context-d1.icon-minus-text-on-apps-with-both.tost5.equivalent\endcsname{0}
\expandafter\def\csname pct@rev.desc.d7.sspro.tau-T-by-type.p2.context-d1.icon-minus-text-on-apps-with-both.tost5.equivalent\endcsname{0.0}
\expandafter\def\csname ppt@rev.desc.d7.sspro.tau-T-by-type.p2.context-d1.icon-minus-text-on-apps-with-both.tost5.equivalent\endcsname{0.0}
\expandafter\def\csname val@rev.desc.d7.sspro.tau-T-by-type.p2.context-d1.icon-minus-text-on-apps-with-both.tost5.p\endcsname{0.757}
\expandafter\def\csname cnt@rev.desc.d7.sspro.tau-T-by-type.p2.context-d1.icon-minus-text-on-apps-with-both.tost5.p\endcsname{1}
\expandafter\def\csname pct@rev.desc.d7.sspro.tau-T-by-type.p2.context-d1.icon-minus-text-on-apps-with-both.tost5.p\endcsname{75.7}
\expandafter\def\csname ppt@rev.desc.d7.sspro.tau-T-by-type.p2.context-d1.icon-minus-text-on-apps-with-both.tost5.p\endcsname{$+$75.7}
\expandafter\def\csname val@rev.desc.d7.sspro.tau-T-by-type.p2.context-d1.icon.acc-F\endcsname{0.714}
\expandafter\def\csname cnt@rev.desc.d7.sspro.tau-T-by-type.p2.context-d1.icon.acc-F\endcsname{1}
\expandafter\def\csname pct@rev.desc.d7.sspro.tau-T-by-type.p2.context-d1.icon.acc-F\endcsname{71.4}
\expandafter\def\csname ppt@rev.desc.d7.sspro.tau-T-by-type.p2.context-d1.icon.acc-F\endcsname{$+$71.4}
\expandafter\def\csname val@rev.desc.d7.sspro.tau-T-by-type.p2.context-d1.icon.acc-direct\endcsname{0.571}
\expandafter\def\csname cnt@rev.desc.d7.sspro.tau-T-by-type.p2.context-d1.icon.acc-direct\endcsname{1}
\expandafter\def\csname pct@rev.desc.d7.sspro.tau-T-by-type.p2.context-d1.icon.acc-direct\endcsname{57.1}
\expandafter\def\csname ppt@rev.desc.d7.sspro.tau-T-by-type.p2.context-d1.icon.acc-direct\endcsname{$+$57.1}
\expandafter\def\csname val@rev.desc.d7.sspro.tau-T-by-type.p2.context-d1.icon.n-apps\endcsname{10}
\expandafter\def\csname cnt@rev.desc.d7.sspro.tau-T-by-type.p2.context-d1.icon.n-apps\endcsname{10}
\expandafter\def\csname pct@rev.desc.d7.sspro.tau-T-by-type.p2.context-d1.icon.n-apps\endcsname{1000.0}
\expandafter\def\csname ppt@rev.desc.d7.sspro.tau-T-by-type.p2.context-d1.icon.n-apps\endcsname{$+$1000.0}
\expandafter\def\csname val@rev.desc.d7.sspro.tau-T-by-type.p2.context-d1.icon.n-tasks\endcsname{28}
\expandafter\def\csname cnt@rev.desc.d7.sspro.tau-T-by-type.p2.context-d1.icon.n-tasks\endcsname{28}
\expandafter\def\csname pct@rev.desc.d7.sspro.tau-T-by-type.p2.context-d1.icon.n-tasks\endcsname{2800.0}
\expandafter\def\csname ppt@rev.desc.d7.sspro.tau-T-by-type.p2.context-d1.icon.n-tasks\endcsname{$+$2800.0}
\expandafter\def\csname val@rev.desc.d7.sspro.tau-T-by-type.p2.context-d1.icon.tau.ci90.0\endcsname{-0.010}
\expandafter\def\csname cnt@rev.desc.d7.sspro.tau-T-by-type.p2.context-d1.icon.tau.ci90.0\endcsname{0}
\expandafter\def\csname pct@rev.desc.d7.sspro.tau-T-by-type.p2.context-d1.icon.tau.ci90.0\endcsname{$-$1.0}
\expandafter\def\csname ppt@rev.desc.d7.sspro.tau-T-by-type.p2.context-d1.icon.tau.ci90.0\endcsname{$-$1.0}
\expandafter\def\csname val@rev.desc.d7.sspro.tau-T-by-type.p2.context-d1.icon.tau.ci90.1\endcsname{0.387}
\expandafter\def\csname cnt@rev.desc.d7.sspro.tau-T-by-type.p2.context-d1.icon.tau.ci90.1\endcsname{0}
\expandafter\def\csname pct@rev.desc.d7.sspro.tau-T-by-type.p2.context-d1.icon.tau.ci90.1\endcsname{38.7}
\expandafter\def\csname ppt@rev.desc.d7.sspro.tau-T-by-type.p2.context-d1.icon.tau.ci90.1\endcsname{$+$38.7}
\expandafter\def\csname val@rev.desc.d7.sspro.tau-T-by-type.p2.context-d1.icon.tau.ci95-excludes-0\endcsname{0}
\expandafter\def\csname cnt@rev.desc.d7.sspro.tau-T-by-type.p2.context-d1.icon.tau.ci95-excludes-0\endcsname{0}
\expandafter\def\csname pct@rev.desc.d7.sspro.tau-T-by-type.p2.context-d1.icon.tau.ci95-excludes-0\endcsname{0.0}
\expandafter\def\csname ppt@rev.desc.d7.sspro.tau-T-by-type.p2.context-d1.icon.tau.ci95-excludes-0\endcsname{0.0}
\expandafter\def\csname val@rev.desc.d7.sspro.tau-T-by-type.p2.context-d1.icon.tau.ci95.0\endcsname{-0.047}
\expandafter\def\csname cnt@rev.desc.d7.sspro.tau-T-by-type.p2.context-d1.icon.tau.ci95.0\endcsname{0}
\expandafter\def\csname pct@rev.desc.d7.sspro.tau-T-by-type.p2.context-d1.icon.tau.ci95.0\endcsname{$-$4.7}
\expandafter\def\csname ppt@rev.desc.d7.sspro.tau-T-by-type.p2.context-d1.icon.tau.ci95.0\endcsname{$-$4.7}
\expandafter\def\csname val@rev.desc.d7.sspro.tau-T-by-type.p2.context-d1.icon.tau.ci95.1\endcsname{0.430}
\expandafter\def\csname cnt@rev.desc.d7.sspro.tau-T-by-type.p2.context-d1.icon.tau.ci95.1\endcsname{0}
\expandafter\def\csname pct@rev.desc.d7.sspro.tau-T-by-type.p2.context-d1.icon.tau.ci95.1\endcsname{43.0}
\expandafter\def\csname ppt@rev.desc.d7.sspro.tau-T-by-type.p2.context-d1.icon.tau.ci95.1\endcsname{$+$43.0}
\expandafter\def\csname val@rev.desc.d7.sspro.tau-T-by-type.p2.context-d1.icon.tau.mean\endcsname{0.187}
\expandafter\def\csname cnt@rev.desc.d7.sspro.tau-T-by-type.p2.context-d1.icon.tau.mean\endcsname{0}
\expandafter\def\csname pct@rev.desc.d7.sspro.tau-T-by-type.p2.context-d1.icon.tau.mean\endcsname{18.7}
\expandafter\def\csname ppt@rev.desc.d7.sspro.tau-T-by-type.p2.context-d1.icon.tau.mean\endcsname{$+$18.7}
\expandafter\def\csname val@rev.desc.d7.sspro.tau-T-by-type.p2.context-d1.icon.tau.n-apps\endcsname{10}
\expandafter\def\csname cnt@rev.desc.d7.sspro.tau-T-by-type.p2.context-d1.icon.tau.n-apps\endcsname{10}
\expandafter\def\csname pct@rev.desc.d7.sspro.tau-T-by-type.p2.context-d1.icon.tau.n-apps\endcsname{1000.0}
\expandafter\def\csname ppt@rev.desc.d7.sspro.tau-T-by-type.p2.context-d1.icon.tau.n-apps\endcsname{$+$1000.0}
\expandafter\def\csname val@rev.desc.d7.sspro.tau-T-by-type.p2.context-d1.icon.tau.tost1.ci90-inside\endcsname{0}
\expandafter\def\csname cnt@rev.desc.d7.sspro.tau-T-by-type.p2.context-d1.icon.tau.tost1.ci90-inside\endcsname{0}
\expandafter\def\csname pct@rev.desc.d7.sspro.tau-T-by-type.p2.context-d1.icon.tau.tost1.ci90-inside\endcsname{0.0}
\expandafter\def\csname ppt@rev.desc.d7.sspro.tau-T-by-type.p2.context-d1.icon.tau.tost1.ci90-inside\endcsname{0.0}
\expandafter\def\csname val@rev.desc.d7.sspro.tau-T-by-type.p2.context-d1.icon.tau.tost1.equivalent\endcsname{0}
\expandafter\def\csname cnt@rev.desc.d7.sspro.tau-T-by-type.p2.context-d1.icon.tau.tost1.equivalent\endcsname{0}
\expandafter\def\csname pct@rev.desc.d7.sspro.tau-T-by-type.p2.context-d1.icon.tau.tost1.equivalent\endcsname{0.0}
\expandafter\def\csname ppt@rev.desc.d7.sspro.tau-T-by-type.p2.context-d1.icon.tau.tost1.equivalent\endcsname{0.0}
\expandafter\def\csname val@rev.desc.d7.sspro.tau-T-by-type.p2.context-d1.icon.tau.tost1.p\endcsname{0.929}
\expandafter\def\csname cnt@rev.desc.d7.sspro.tau-T-by-type.p2.context-d1.icon.tau.tost1.p\endcsname{1}
\expandafter\def\csname pct@rev.desc.d7.sspro.tau-T-by-type.p2.context-d1.icon.tau.tost1.p\endcsname{92.9}
\expandafter\def\csname ppt@rev.desc.d7.sspro.tau-T-by-type.p2.context-d1.icon.tau.tost1.p\endcsname{$+$92.9}
\expandafter\def\csname val@rev.desc.d7.sspro.tau-T-by-type.p2.context-d1.icon.tau.tost5.ci90-inside\endcsname{0}
\expandafter\def\csname cnt@rev.desc.d7.sspro.tau-T-by-type.p2.context-d1.icon.tau.tost5.ci90-inside\endcsname{0}
\expandafter\def\csname pct@rev.desc.d7.sspro.tau-T-by-type.p2.context-d1.icon.tau.tost5.ci90-inside\endcsname{0.0}
\expandafter\def\csname ppt@rev.desc.d7.sspro.tau-T-by-type.p2.context-d1.icon.tau.tost5.ci90-inside\endcsname{0.0}
\expandafter\def\csname val@rev.desc.d7.sspro.tau-T-by-type.p2.context-d1.icon.tau.tost5.equivalent\endcsname{0}
\expandafter\def\csname cnt@rev.desc.d7.sspro.tau-T-by-type.p2.context-d1.icon.tau.tost5.equivalent\endcsname{0}
\expandafter\def\csname pct@rev.desc.d7.sspro.tau-T-by-type.p2.context-d1.icon.tau.tost5.equivalent\endcsname{0.0}
\expandafter\def\csname ppt@rev.desc.d7.sspro.tau-T-by-type.p2.context-d1.icon.tau.tost5.equivalent\endcsname{0.0}
\expandafter\def\csname val@rev.desc.d7.sspro.tau-T-by-type.p2.context-d1.icon.tau.tost5.p\endcsname{0.885}
\expandafter\def\csname cnt@rev.desc.d7.sspro.tau-T-by-type.p2.context-d1.icon.tau.tost5.p\endcsname{1}
\expandafter\def\csname pct@rev.desc.d7.sspro.tau-T-by-type.p2.context-d1.icon.tau.tost5.p\endcsname{88.5}
\expandafter\def\csname ppt@rev.desc.d7.sspro.tau-T-by-type.p2.context-d1.icon.tau.tost5.p\endcsname{$+$88.5}
\expandafter\def\csname val@rev.desc.d7.sspro.tau-T-by-type.p2.context-d1.text.acc-F\endcsname{0.956}
\expandafter\def\csname cnt@rev.desc.d7.sspro.tau-T-by-type.p2.context-d1.text.acc-F\endcsname{1}
\expandafter\def\csname pct@rev.desc.d7.sspro.tau-T-by-type.p2.context-d1.text.acc-F\endcsname{95.6}
\expandafter\def\csname ppt@rev.desc.d7.sspro.tau-T-by-type.p2.context-d1.text.acc-F\endcsname{$+$95.6}
\expandafter\def\csname val@rev.desc.d7.sspro.tau-T-by-type.p2.context-d1.text.acc-direct\endcsname{0.838}
\expandafter\def\csname cnt@rev.desc.d7.sspro.tau-T-by-type.p2.context-d1.text.acc-direct\endcsname{1}
\expandafter\def\csname pct@rev.desc.d7.sspro.tau-T-by-type.p2.context-d1.text.acc-direct\endcsname{83.8}
\expandafter\def\csname ppt@rev.desc.d7.sspro.tau-T-by-type.p2.context-d1.text.acc-direct\endcsname{$+$83.8}
\expandafter\def\csname val@rev.desc.d7.sspro.tau-T-by-type.p2.context-d1.text.n-apps\endcsname{12}
\expandafter\def\csname cnt@rev.desc.d7.sspro.tau-T-by-type.p2.context-d1.text.n-apps\endcsname{12}
\expandafter\def\csname pct@rev.desc.d7.sspro.tau-T-by-type.p2.context-d1.text.n-apps\endcsname{1200.0}
\expandafter\def\csname ppt@rev.desc.d7.sspro.tau-T-by-type.p2.context-d1.text.n-apps\endcsname{$+$1200.0}
\expandafter\def\csname val@rev.desc.d7.sspro.tau-T-by-type.p2.context-d1.text.n-tasks\endcsname{68}
\expandafter\def\csname cnt@rev.desc.d7.sspro.tau-T-by-type.p2.context-d1.text.n-tasks\endcsname{68}
\expandafter\def\csname pct@rev.desc.d7.sspro.tau-T-by-type.p2.context-d1.text.n-tasks\endcsname{6800.0}
\expandafter\def\csname ppt@rev.desc.d7.sspro.tau-T-by-type.p2.context-d1.text.n-tasks\endcsname{$+$6800.0}
\expandafter\def\csname val@rev.desc.d7.sspro.tau-T-by-type.p2.context-d1.text.tau.ci90.0\endcsname{0.034}
\expandafter\def\csname cnt@rev.desc.d7.sspro.tau-T-by-type.p2.context-d1.text.tau.ci90.0\endcsname{0}
\expandafter\def\csname pct@rev.desc.d7.sspro.tau-T-by-type.p2.context-d1.text.tau.ci90.0\endcsname{3.4}
\expandafter\def\csname ppt@rev.desc.d7.sspro.tau-T-by-type.p2.context-d1.text.tau.ci90.0\endcsname{$+$3.4}
\expandafter\def\csname val@rev.desc.d7.sspro.tau-T-by-type.p2.context-d1.text.tau.ci90.1\endcsname{0.163}
\expandafter\def\csname cnt@rev.desc.d7.sspro.tau-T-by-type.p2.context-d1.text.tau.ci90.1\endcsname{0}
\expandafter\def\csname pct@rev.desc.d7.sspro.tau-T-by-type.p2.context-d1.text.tau.ci90.1\endcsname{16.3}
\expandafter\def\csname ppt@rev.desc.d7.sspro.tau-T-by-type.p2.context-d1.text.tau.ci90.1\endcsname{$+$16.3}
\expandafter\def\csname val@rev.desc.d7.sspro.tau-T-by-type.p2.context-d1.text.tau.ci95-excludes-0\endcsname{1}
\expandafter\def\csname cnt@rev.desc.d7.sspro.tau-T-by-type.p2.context-d1.text.tau.ci95-excludes-0\endcsname{1}
\expandafter\def\csname pct@rev.desc.d7.sspro.tau-T-by-type.p2.context-d1.text.tau.ci95-excludes-0\endcsname{100.0}
\expandafter\def\csname ppt@rev.desc.d7.sspro.tau-T-by-type.p2.context-d1.text.tau.ci95-excludes-0\endcsname{$+$100.0}
\expandafter\def\csname val@rev.desc.d7.sspro.tau-T-by-type.p2.context-d1.text.tau.ci95.0\endcsname{0.024}
\expandafter\def\csname cnt@rev.desc.d7.sspro.tau-T-by-type.p2.context-d1.text.tau.ci95.0\endcsname{0}
\expandafter\def\csname pct@rev.desc.d7.sspro.tau-T-by-type.p2.context-d1.text.tau.ci95.0\endcsname{2.4}
\expandafter\def\csname ppt@rev.desc.d7.sspro.tau-T-by-type.p2.context-d1.text.tau.ci95.0\endcsname{$+$2.4}
\expandafter\def\csname val@rev.desc.d7.sspro.tau-T-by-type.p2.context-d1.text.tau.ci95.1\endcsname{0.181}
\expandafter\def\csname cnt@rev.desc.d7.sspro.tau-T-by-type.p2.context-d1.text.tau.ci95.1\endcsname{0}
\expandafter\def\csname pct@rev.desc.d7.sspro.tau-T-by-type.p2.context-d1.text.tau.ci95.1\endcsname{18.1}
\expandafter\def\csname ppt@rev.desc.d7.sspro.tau-T-by-type.p2.context-d1.text.tau.ci95.1\endcsname{$+$18.1}
\expandafter\def\csname val@rev.desc.d7.sspro.tau-T-by-type.p2.context-d1.text.tau.mean\endcsname{0.092}
\expandafter\def\csname cnt@rev.desc.d7.sspro.tau-T-by-type.p2.context-d1.text.tau.mean\endcsname{0}
\expandafter\def\csname pct@rev.desc.d7.sspro.tau-T-by-type.p2.context-d1.text.tau.mean\endcsname{9.2}
\expandafter\def\csname ppt@rev.desc.d7.sspro.tau-T-by-type.p2.context-d1.text.tau.mean\endcsname{$+$9.2}
\expandafter\def\csname val@rev.desc.d7.sspro.tau-T-by-type.p2.context-d1.text.tau.n-apps\endcsname{12}
\expandafter\def\csname cnt@rev.desc.d7.sspro.tau-T-by-type.p2.context-d1.text.tau.n-apps\endcsname{12}
\expandafter\def\csname pct@rev.desc.d7.sspro.tau-T-by-type.p2.context-d1.text.tau.n-apps\endcsname{1200.0}
\expandafter\def\csname ppt@rev.desc.d7.sspro.tau-T-by-type.p2.context-d1.text.tau.n-apps\endcsname{$+$1200.0}
\expandafter\def\csname val@rev.desc.d7.sspro.tau-T-by-type.p2.context-d1.text.tau.tost1.ci90-inside\endcsname{0}
\expandafter\def\csname cnt@rev.desc.d7.sspro.tau-T-by-type.p2.context-d1.text.tau.tost1.ci90-inside\endcsname{0}
\expandafter\def\csname pct@rev.desc.d7.sspro.tau-T-by-type.p2.context-d1.text.tau.tost1.ci90-inside\endcsname{0.0}
\expandafter\def\csname ppt@rev.desc.d7.sspro.tau-T-by-type.p2.context-d1.text.tau.tost1.ci90-inside\endcsname{0.0}
\expandafter\def\csname val@rev.desc.d7.sspro.tau-T-by-type.p2.context-d1.text.tau.tost1.equivalent\endcsname{0}
\expandafter\def\csname cnt@rev.desc.d7.sspro.tau-T-by-type.p2.context-d1.text.tau.tost1.equivalent\endcsname{0}
\expandafter\def\csname pct@rev.desc.d7.sspro.tau-T-by-type.p2.context-d1.text.tau.tost1.equivalent\endcsname{0.0}
\expandafter\def\csname ppt@rev.desc.d7.sspro.tau-T-by-type.p2.context-d1.text.tau.tost1.equivalent\endcsname{0.0}
\expandafter\def\csname val@rev.desc.d7.sspro.tau-T-by-type.p2.context-d1.text.tau.tost1.p\endcsname{0.999}
\expandafter\def\csname cnt@rev.desc.d7.sspro.tau-T-by-type.p2.context-d1.text.tau.tost1.p\endcsname{1}
\expandafter\def\csname pct@rev.desc.d7.sspro.tau-T-by-type.p2.context-d1.text.tau.tost1.p\endcsname{99.9}
\expandafter\def\csname ppt@rev.desc.d7.sspro.tau-T-by-type.p2.context-d1.text.tau.tost1.p\endcsname{$+$99.9}
\expandafter\def\csname val@rev.desc.d7.sspro.tau-T-by-type.p2.context-d1.text.tau.tost5.ci90-inside\endcsname{0}
\expandafter\def\csname cnt@rev.desc.d7.sspro.tau-T-by-type.p2.context-d1.text.tau.tost5.ci90-inside\endcsname{0}
\expandafter\def\csname pct@rev.desc.d7.sspro.tau-T-by-type.p2.context-d1.text.tau.tost5.ci90-inside\endcsname{0.0}
\expandafter\def\csname ppt@rev.desc.d7.sspro.tau-T-by-type.p2.context-d1.text.tau.tost5.ci90-inside\endcsname{0.0}
\expandafter\def\csname val@rev.desc.d7.sspro.tau-T-by-type.p2.context-d1.text.tau.tost5.equivalent\endcsname{0}
\expandafter\def\csname cnt@rev.desc.d7.sspro.tau-T-by-type.p2.context-d1.text.tau.tost5.equivalent\endcsname{0}
\expandafter\def\csname pct@rev.desc.d7.sspro.tau-T-by-type.p2.context-d1.text.tau.tost5.equivalent\endcsname{0.0}
\expandafter\def\csname ppt@rev.desc.d7.sspro.tau-T-by-type.p2.context-d1.text.tau.tost5.equivalent\endcsname{0.0}
\expandafter\def\csname val@rev.desc.d7.sspro.tau-T-by-type.p2.context-d1.text.tau.tost5.p\endcsname{0.854}
\expandafter\def\csname cnt@rev.desc.d7.sspro.tau-T-by-type.p2.context-d1.text.tau.tost5.p\endcsname{1}
\expandafter\def\csname pct@rev.desc.d7.sspro.tau-T-by-type.p2.context-d1.text.tau.tost5.p\endcsname{85.4}
\expandafter\def\csname ppt@rev.desc.d7.sspro.tau-T-by-type.p2.context-d1.text.tau.tost5.p\endcsname{$+$85.4}
\expandafter\def\csname val@rev.desc.d7.sspro.tau-T-by-type.p2.dual-d2.icon-minus-text-on-apps-with-both.ci90.0\endcsname{0.020}
\expandafter\def\csname cnt@rev.desc.d7.sspro.tau-T-by-type.p2.dual-d2.icon-minus-text-on-apps-with-both.ci90.0\endcsname{0}
\expandafter\def\csname pct@rev.desc.d7.sspro.tau-T-by-type.p2.dual-d2.icon-minus-text-on-apps-with-both.ci90.0\endcsname{2.0}
\expandafter\def\csname ppt@rev.desc.d7.sspro.tau-T-by-type.p2.dual-d2.icon-minus-text-on-apps-with-both.ci90.0\endcsname{$+$2.0}
\expandafter\def\csname val@rev.desc.d7.sspro.tau-T-by-type.p2.dual-d2.icon-minus-text-on-apps-with-both.ci90.1\endcsname{0.413}
\expandafter\def\csname cnt@rev.desc.d7.sspro.tau-T-by-type.p2.dual-d2.icon-minus-text-on-apps-with-both.ci90.1\endcsname{0}
\expandafter\def\csname pct@rev.desc.d7.sspro.tau-T-by-type.p2.dual-d2.icon-minus-text-on-apps-with-both.ci90.1\endcsname{41.3}
\expandafter\def\csname ppt@rev.desc.d7.sspro.tau-T-by-type.p2.dual-d2.icon-minus-text-on-apps-with-both.ci90.1\endcsname{$+$41.3}
\expandafter\def\csname val@rev.desc.d7.sspro.tau-T-by-type.p2.dual-d2.icon-minus-text-on-apps-with-both.ci95-excludes-0\endcsname{0}
\expandafter\def\csname cnt@rev.desc.d7.sspro.tau-T-by-type.p2.dual-d2.icon-minus-text-on-apps-with-both.ci95-excludes-0\endcsname{0}
\expandafter\def\csname pct@rev.desc.d7.sspro.tau-T-by-type.p2.dual-d2.icon-minus-text-on-apps-with-both.ci95-excludes-0\endcsname{0.0}
\expandafter\def\csname ppt@rev.desc.d7.sspro.tau-T-by-type.p2.dual-d2.icon-minus-text-on-apps-with-both.ci95-excludes-0\endcsname{0.0}
\expandafter\def\csname val@rev.desc.d7.sspro.tau-T-by-type.p2.dual-d2.icon-minus-text-on-apps-with-both.ci95.0\endcsname{-0.005}
\expandafter\def\csname cnt@rev.desc.d7.sspro.tau-T-by-type.p2.dual-d2.icon-minus-text-on-apps-with-both.ci95.0\endcsname{0}
\expandafter\def\csname pct@rev.desc.d7.sspro.tau-T-by-type.p2.dual-d2.icon-minus-text-on-apps-with-both.ci95.0\endcsname{$-$0.5}
\expandafter\def\csname ppt@rev.desc.d7.sspro.tau-T-by-type.p2.dual-d2.icon-minus-text-on-apps-with-both.ci95.0\endcsname{$-$0.5}
\expandafter\def\csname val@rev.desc.d7.sspro.tau-T-by-type.p2.dual-d2.icon-minus-text-on-apps-with-both.ci95.1\endcsname{0.447}
\expandafter\def\csname cnt@rev.desc.d7.sspro.tau-T-by-type.p2.dual-d2.icon-minus-text-on-apps-with-both.ci95.1\endcsname{0}
\expandafter\def\csname pct@rev.desc.d7.sspro.tau-T-by-type.p2.dual-d2.icon-minus-text-on-apps-with-both.ci95.1\endcsname{44.7}
\expandafter\def\csname ppt@rev.desc.d7.sspro.tau-T-by-type.p2.dual-d2.icon-minus-text-on-apps-with-both.ci95.1\endcsname{$+$44.7}
\expandafter\def\csname val@rev.desc.d7.sspro.tau-T-by-type.p2.dual-d2.icon-minus-text-on-apps-with-both.mean\endcsname{0.211}
\expandafter\def\csname cnt@rev.desc.d7.sspro.tau-T-by-type.p2.dual-d2.icon-minus-text-on-apps-with-both.mean\endcsname{0}
\expandafter\def\csname pct@rev.desc.d7.sspro.tau-T-by-type.p2.dual-d2.icon-minus-text-on-apps-with-both.mean\endcsname{21.1}
\expandafter\def\csname ppt@rev.desc.d7.sspro.tau-T-by-type.p2.dual-d2.icon-minus-text-on-apps-with-both.mean\endcsname{$+$21.1}
\expandafter\def\csname val@rev.desc.d7.sspro.tau-T-by-type.p2.dual-d2.icon-minus-text-on-apps-with-both.n-apps\endcsname{10}
\expandafter\def\csname cnt@rev.desc.d7.sspro.tau-T-by-type.p2.dual-d2.icon-minus-text-on-apps-with-both.n-apps\endcsname{10}
\expandafter\def\csname pct@rev.desc.d7.sspro.tau-T-by-type.p2.dual-d2.icon-minus-text-on-apps-with-both.n-apps\endcsname{1000.0}
\expandafter\def\csname ppt@rev.desc.d7.sspro.tau-T-by-type.p2.dual-d2.icon-minus-text-on-apps-with-both.n-apps\endcsname{$+$1000.0}
\expandafter\def\csname val@rev.desc.d7.sspro.tau-T-by-type.p2.dual-d2.icon-minus-text-on-apps-with-both.tost1.ci90-inside\endcsname{0}
\expandafter\def\csname cnt@rev.desc.d7.sspro.tau-T-by-type.p2.dual-d2.icon-minus-text-on-apps-with-both.tost1.ci90-inside\endcsname{0}
\expandafter\def\csname pct@rev.desc.d7.sspro.tau-T-by-type.p2.dual-d2.icon-minus-text-on-apps-with-both.tost1.ci90-inside\endcsname{0.0}
\expandafter\def\csname ppt@rev.desc.d7.sspro.tau-T-by-type.p2.dual-d2.icon-minus-text-on-apps-with-both.tost1.ci90-inside\endcsname{0.0}
\expandafter\def\csname val@rev.desc.d7.sspro.tau-T-by-type.p2.dual-d2.icon-minus-text-on-apps-with-both.tost1.equivalent\endcsname{0}
\expandafter\def\csname cnt@rev.desc.d7.sspro.tau-T-by-type.p2.dual-d2.icon-minus-text-on-apps-with-both.tost1.equivalent\endcsname{0}
\expandafter\def\csname pct@rev.desc.d7.sspro.tau-T-by-type.p2.dual-d2.icon-minus-text-on-apps-with-both.tost1.equivalent\endcsname{0.0}
\expandafter\def\csname ppt@rev.desc.d7.sspro.tau-T-by-type.p2.dual-d2.icon-minus-text-on-apps-with-both.tost1.equivalent\endcsname{0.0}
\expandafter\def\csname val@rev.desc.d7.sspro.tau-T-by-type.p2.dual-d2.icon-minus-text-on-apps-with-both.tost1.p\endcsname{0.962}
\expandafter\def\csname cnt@rev.desc.d7.sspro.tau-T-by-type.p2.dual-d2.icon-minus-text-on-apps-with-both.tost1.p\endcsname{1}
\expandafter\def\csname pct@rev.desc.d7.sspro.tau-T-by-type.p2.dual-d2.icon-minus-text-on-apps-with-both.tost1.p\endcsname{96.2}
\expandafter\def\csname ppt@rev.desc.d7.sspro.tau-T-by-type.p2.dual-d2.icon-minus-text-on-apps-with-both.tost1.p\endcsname{$+$96.2}
\expandafter\def\csname val@rev.desc.d7.sspro.tau-T-by-type.p2.dual-d2.icon-minus-text-on-apps-with-both.tost5.ci90-inside\endcsname{0}
\expandafter\def\csname cnt@rev.desc.d7.sspro.tau-T-by-type.p2.dual-d2.icon-minus-text-on-apps-with-both.tost5.ci90-inside\endcsname{0}
\expandafter\def\csname pct@rev.desc.d7.sspro.tau-T-by-type.p2.dual-d2.icon-minus-text-on-apps-with-both.tost5.ci90-inside\endcsname{0.0}
\expandafter\def\csname ppt@rev.desc.d7.sspro.tau-T-by-type.p2.dual-d2.icon-minus-text-on-apps-with-both.tost5.ci90-inside\endcsname{0.0}
\expandafter\def\csname val@rev.desc.d7.sspro.tau-T-by-type.p2.dual-d2.icon-minus-text-on-apps-with-both.tost5.equivalent\endcsname{0}
\expandafter\def\csname cnt@rev.desc.d7.sspro.tau-T-by-type.p2.dual-d2.icon-minus-text-on-apps-with-both.tost5.equivalent\endcsname{0}
\expandafter\def\csname pct@rev.desc.d7.sspro.tau-T-by-type.p2.dual-d2.icon-minus-text-on-apps-with-both.tost5.equivalent\endcsname{0.0}
\expandafter\def\csname ppt@rev.desc.d7.sspro.tau-T-by-type.p2.dual-d2.icon-minus-text-on-apps-with-both.tost5.equivalent\endcsname{0.0}
\expandafter\def\csname val@rev.desc.d7.sspro.tau-T-by-type.p2.dual-d2.icon-minus-text-on-apps-with-both.tost5.p\endcsname{0.912}
\expandafter\def\csname cnt@rev.desc.d7.sspro.tau-T-by-type.p2.dual-d2.icon-minus-text-on-apps-with-both.tost5.p\endcsname{1}
\expandafter\def\csname pct@rev.desc.d7.sspro.tau-T-by-type.p2.dual-d2.icon-minus-text-on-apps-with-both.tost5.p\endcsname{91.2}
\expandafter\def\csname ppt@rev.desc.d7.sspro.tau-T-by-type.p2.dual-d2.icon-minus-text-on-apps-with-both.tost5.p\endcsname{$+$91.2}
\expandafter\def\csname val@rev.desc.d7.sspro.tau-T-by-type.p2.dual-d2.icon.acc-F\endcsname{0.786}
\expandafter\def\csname cnt@rev.desc.d7.sspro.tau-T-by-type.p2.dual-d2.icon.acc-F\endcsname{1}
\expandafter\def\csname pct@rev.desc.d7.sspro.tau-T-by-type.p2.dual-d2.icon.acc-F\endcsname{78.6}
\expandafter\def\csname ppt@rev.desc.d7.sspro.tau-T-by-type.p2.dual-d2.icon.acc-F\endcsname{$+$78.6}
\expandafter\def\csname val@rev.desc.d7.sspro.tau-T-by-type.p2.dual-d2.icon.acc-direct\endcsname{0.571}
\expandafter\def\csname cnt@rev.desc.d7.sspro.tau-T-by-type.p2.dual-d2.icon.acc-direct\endcsname{1}
\expandafter\def\csname pct@rev.desc.d7.sspro.tau-T-by-type.p2.dual-d2.icon.acc-direct\endcsname{57.1}
\expandafter\def\csname ppt@rev.desc.d7.sspro.tau-T-by-type.p2.dual-d2.icon.acc-direct\endcsname{$+$57.1}
\expandafter\def\csname val@rev.desc.d7.sspro.tau-T-by-type.p2.dual-d2.icon.n-apps\endcsname{10}
\expandafter\def\csname cnt@rev.desc.d7.sspro.tau-T-by-type.p2.dual-d2.icon.n-apps\endcsname{10}
\expandafter\def\csname pct@rev.desc.d7.sspro.tau-T-by-type.p2.dual-d2.icon.n-apps\endcsname{1000.0}
\expandafter\def\csname ppt@rev.desc.d7.sspro.tau-T-by-type.p2.dual-d2.icon.n-apps\endcsname{$+$1000.0}
\expandafter\def\csname val@rev.desc.d7.sspro.tau-T-by-type.p2.dual-d2.icon.n-tasks\endcsname{28}
\expandafter\def\csname cnt@rev.desc.d7.sspro.tau-T-by-type.p2.dual-d2.icon.n-tasks\endcsname{28}
\expandafter\def\csname pct@rev.desc.d7.sspro.tau-T-by-type.p2.dual-d2.icon.n-tasks\endcsname{2800.0}
\expandafter\def\csname ppt@rev.desc.d7.sspro.tau-T-by-type.p2.dual-d2.icon.n-tasks\endcsname{$+$2800.0}
\expandafter\def\csname val@rev.desc.d7.sspro.tau-T-by-type.p2.dual-d2.icon.tau.ci90.0\endcsname{0.090}
\expandafter\def\csname cnt@rev.desc.d7.sspro.tau-T-by-type.p2.dual-d2.icon.tau.ci90.0\endcsname{0}
\expandafter\def\csname pct@rev.desc.d7.sspro.tau-T-by-type.p2.dual-d2.icon.tau.ci90.0\endcsname{9.0}
\expandafter\def\csname ppt@rev.desc.d7.sspro.tau-T-by-type.p2.dual-d2.icon.tau.ci90.0\endcsname{$+$9.0}
\expandafter\def\csname val@rev.desc.d7.sspro.tau-T-by-type.p2.dual-d2.icon.tau.ci90.1\endcsname{0.437}
\expandafter\def\csname cnt@rev.desc.d7.sspro.tau-T-by-type.p2.dual-d2.icon.tau.ci90.1\endcsname{0}
\expandafter\def\csname pct@rev.desc.d7.sspro.tau-T-by-type.p2.dual-d2.icon.tau.ci90.1\endcsname{43.7}
\expandafter\def\csname ppt@rev.desc.d7.sspro.tau-T-by-type.p2.dual-d2.icon.tau.ci90.1\endcsname{$+$43.7}
\expandafter\def\csname val@rev.desc.d7.sspro.tau-T-by-type.p2.dual-d2.icon.tau.ci95-excludes-0\endcsname{1}
\expandafter\def\csname cnt@rev.desc.d7.sspro.tau-T-by-type.p2.dual-d2.icon.tau.ci95-excludes-0\endcsname{1}
\expandafter\def\csname pct@rev.desc.d7.sspro.tau-T-by-type.p2.dual-d2.icon.tau.ci95-excludes-0\endcsname{100.0}
\expandafter\def\csname ppt@rev.desc.d7.sspro.tau-T-by-type.p2.dual-d2.icon.tau.ci95-excludes-0\endcsname{$+$100.0}
\expandafter\def\csname val@rev.desc.d7.sspro.tau-T-by-type.p2.dual-d2.icon.tau.ci95.0\endcsname{0.067}
\expandafter\def\csname cnt@rev.desc.d7.sspro.tau-T-by-type.p2.dual-d2.icon.tau.ci95.0\endcsname{0}
\expandafter\def\csname pct@rev.desc.d7.sspro.tau-T-by-type.p2.dual-d2.icon.tau.ci95.0\endcsname{6.7}
\expandafter\def\csname ppt@rev.desc.d7.sspro.tau-T-by-type.p2.dual-d2.icon.tau.ci95.0\endcsname{$+$6.7}
\expandafter\def\csname val@rev.desc.d7.sspro.tau-T-by-type.p2.dual-d2.icon.tau.ci95.1\endcsname{0.480}
\expandafter\def\csname cnt@rev.desc.d7.sspro.tau-T-by-type.p2.dual-d2.icon.tau.ci95.1\endcsname{0}
\expandafter\def\csname pct@rev.desc.d7.sspro.tau-T-by-type.p2.dual-d2.icon.tau.ci95.1\endcsname{48.0}
\expandafter\def\csname ppt@rev.desc.d7.sspro.tau-T-by-type.p2.dual-d2.icon.tau.ci95.1\endcsname{$+$48.0}
\expandafter\def\csname val@rev.desc.d7.sspro.tau-T-by-type.p2.dual-d2.icon.tau.mean\endcsname{0.257}
\expandafter\def\csname cnt@rev.desc.d7.sspro.tau-T-by-type.p2.dual-d2.icon.tau.mean\endcsname{0}
\expandafter\def\csname pct@rev.desc.d7.sspro.tau-T-by-type.p2.dual-d2.icon.tau.mean\endcsname{25.7}
\expandafter\def\csname ppt@rev.desc.d7.sspro.tau-T-by-type.p2.dual-d2.icon.tau.mean\endcsname{$+$25.7}
\expandafter\def\csname val@rev.desc.d7.sspro.tau-T-by-type.p2.dual-d2.icon.tau.n-apps\endcsname{10}
\expandafter\def\csname cnt@rev.desc.d7.sspro.tau-T-by-type.p2.dual-d2.icon.tau.n-apps\endcsname{10}
\expandafter\def\csname pct@rev.desc.d7.sspro.tau-T-by-type.p2.dual-d2.icon.tau.n-apps\endcsname{1000.0}
\expandafter\def\csname ppt@rev.desc.d7.sspro.tau-T-by-type.p2.dual-d2.icon.tau.n-apps\endcsname{$+$1000.0}
\expandafter\def\csname val@rev.desc.d7.sspro.tau-T-by-type.p2.dual-d2.icon.tau.tost1.ci90-inside\endcsname{0}
\expandafter\def\csname cnt@rev.desc.d7.sspro.tau-T-by-type.p2.dual-d2.icon.tau.tost1.ci90-inside\endcsname{0}
\expandafter\def\csname pct@rev.desc.d7.sspro.tau-T-by-type.p2.dual-d2.icon.tau.tost1.ci90-inside\endcsname{0.0}
\expandafter\def\csname ppt@rev.desc.d7.sspro.tau-T-by-type.p2.dual-d2.icon.tau.tost1.ci90-inside\endcsname{0.0}
\expandafter\def\csname val@rev.desc.d7.sspro.tau-T-by-type.p2.dual-d2.icon.tau.tost1.equivalent\endcsname{0}
\expandafter\def\csname cnt@rev.desc.d7.sspro.tau-T-by-type.p2.dual-d2.icon.tau.tost1.equivalent\endcsname{0}
\expandafter\def\csname pct@rev.desc.d7.sspro.tau-T-by-type.p2.dual-d2.icon.tau.tost1.equivalent\endcsname{0.0}
\expandafter\def\csname ppt@rev.desc.d7.sspro.tau-T-by-type.p2.dual-d2.icon.tau.tost1.equivalent\endcsname{0.0}
\expandafter\def\csname val@rev.desc.d7.sspro.tau-T-by-type.p2.dual-d2.icon.tau.tost1.p\endcsname{0.998}
\expandafter\def\csname cnt@rev.desc.d7.sspro.tau-T-by-type.p2.dual-d2.icon.tau.tost1.p\endcsname{1}
\expandafter\def\csname pct@rev.desc.d7.sspro.tau-T-by-type.p2.dual-d2.icon.tau.tost1.p\endcsname{99.8}
\expandafter\def\csname ppt@rev.desc.d7.sspro.tau-T-by-type.p2.dual-d2.icon.tau.tost1.p\endcsname{$+$99.8}
\expandafter\def\csname val@rev.desc.d7.sspro.tau-T-by-type.p2.dual-d2.icon.tau.tost5.ci90-inside\endcsname{0}
\expandafter\def\csname cnt@rev.desc.d7.sspro.tau-T-by-type.p2.dual-d2.icon.tau.tost5.ci90-inside\endcsname{0}
\expandafter\def\csname pct@rev.desc.d7.sspro.tau-T-by-type.p2.dual-d2.icon.tau.tost5.ci90-inside\endcsname{0.0}
\expandafter\def\csname ppt@rev.desc.d7.sspro.tau-T-by-type.p2.dual-d2.icon.tau.tost5.ci90-inside\endcsname{0.0}
\expandafter\def\csname val@rev.desc.d7.sspro.tau-T-by-type.p2.dual-d2.icon.tau.tost5.equivalent\endcsname{0}
\expandafter\def\csname cnt@rev.desc.d7.sspro.tau-T-by-type.p2.dual-d2.icon.tau.tost5.equivalent\endcsname{0}
\expandafter\def\csname pct@rev.desc.d7.sspro.tau-T-by-type.p2.dual-d2.icon.tau.tost5.equivalent\endcsname{0.0}
\expandafter\def\csname ppt@rev.desc.d7.sspro.tau-T-by-type.p2.dual-d2.icon.tau.tost5.equivalent\endcsname{0.0}
\expandafter\def\csname val@rev.desc.d7.sspro.tau-T-by-type.p2.dual-d2.icon.tau.tost5.p\endcsname{0.987}
\expandafter\def\csname cnt@rev.desc.d7.sspro.tau-T-by-type.p2.dual-d2.icon.tau.tost5.p\endcsname{1}
\expandafter\def\csname pct@rev.desc.d7.sspro.tau-T-by-type.p2.dual-d2.icon.tau.tost5.p\endcsname{98.7}
\expandafter\def\csname ppt@rev.desc.d7.sspro.tau-T-by-type.p2.dual-d2.icon.tau.tost5.p\endcsname{$+$98.7}
\expandafter\def\csname val@rev.desc.d7.sspro.tau-T-by-type.p2.dual-d2.text.acc-F\endcsname{0.971}
\expandafter\def\csname cnt@rev.desc.d7.sspro.tau-T-by-type.p2.dual-d2.text.acc-F\endcsname{1}
\expandafter\def\csname pct@rev.desc.d7.sspro.tau-T-by-type.p2.dual-d2.text.acc-F\endcsname{97.1}
\expandafter\def\csname ppt@rev.desc.d7.sspro.tau-T-by-type.p2.dual-d2.text.acc-F\endcsname{$+$97.1}
\expandafter\def\csname val@rev.desc.d7.sspro.tau-T-by-type.p2.dual-d2.text.acc-direct\endcsname{0.838}
\expandafter\def\csname cnt@rev.desc.d7.sspro.tau-T-by-type.p2.dual-d2.text.acc-direct\endcsname{1}
\expandafter\def\csname pct@rev.desc.d7.sspro.tau-T-by-type.p2.dual-d2.text.acc-direct\endcsname{83.8}
\expandafter\def\csname ppt@rev.desc.d7.sspro.tau-T-by-type.p2.dual-d2.text.acc-direct\endcsname{$+$83.8}
\expandafter\def\csname val@rev.desc.d7.sspro.tau-T-by-type.p2.dual-d2.text.n-apps\endcsname{12}
\expandafter\def\csname cnt@rev.desc.d7.sspro.tau-T-by-type.p2.dual-d2.text.n-apps\endcsname{12}
\expandafter\def\csname pct@rev.desc.d7.sspro.tau-T-by-type.p2.dual-d2.text.n-apps\endcsname{1200.0}
\expandafter\def\csname ppt@rev.desc.d7.sspro.tau-T-by-type.p2.dual-d2.text.n-apps\endcsname{$+$1200.0}
\expandafter\def\csname val@rev.desc.d7.sspro.tau-T-by-type.p2.dual-d2.text.n-tasks\endcsname{68}
\expandafter\def\csname cnt@rev.desc.d7.sspro.tau-T-by-type.p2.dual-d2.text.n-tasks\endcsname{68}
\expandafter\def\csname pct@rev.desc.d7.sspro.tau-T-by-type.p2.dual-d2.text.n-tasks\endcsname{6800.0}
\expandafter\def\csname ppt@rev.desc.d7.sspro.tau-T-by-type.p2.dual-d2.text.n-tasks\endcsname{$+$6800.0}
\expandafter\def\csname val@rev.desc.d7.sspro.tau-T-by-type.p2.dual-d2.text.tau.ci90.0\endcsname{0.031}
\expandafter\def\csname cnt@rev.desc.d7.sspro.tau-T-by-type.p2.dual-d2.text.tau.ci90.0\endcsname{0}
\expandafter\def\csname pct@rev.desc.d7.sspro.tau-T-by-type.p2.dual-d2.text.tau.ci90.0\endcsname{3.1}
\expandafter\def\csname ppt@rev.desc.d7.sspro.tau-T-by-type.p2.dual-d2.text.tau.ci90.0\endcsname{$+$3.1}
\expandafter\def\csname val@rev.desc.d7.sspro.tau-T-by-type.p2.dual-d2.text.tau.ci90.1\endcsname{0.173}
\expandafter\def\csname cnt@rev.desc.d7.sspro.tau-T-by-type.p2.dual-d2.text.tau.ci90.1\endcsname{0}
\expandafter\def\csname pct@rev.desc.d7.sspro.tau-T-by-type.p2.dual-d2.text.tau.ci90.1\endcsname{17.3}
\expandafter\def\csname ppt@rev.desc.d7.sspro.tau-T-by-type.p2.dual-d2.text.tau.ci90.1\endcsname{$+$17.3}
\expandafter\def\csname val@rev.desc.d7.sspro.tau-T-by-type.p2.dual-d2.text.tau.ci95-excludes-0\endcsname{1}
\expandafter\def\csname cnt@rev.desc.d7.sspro.tau-T-by-type.p2.dual-d2.text.tau.ci95-excludes-0\endcsname{1}
\expandafter\def\csname pct@rev.desc.d7.sspro.tau-T-by-type.p2.dual-d2.text.tau.ci95-excludes-0\endcsname{100.0}
\expandafter\def\csname ppt@rev.desc.d7.sspro.tau-T-by-type.p2.dual-d2.text.tau.ci95-excludes-0\endcsname{$+$100.0}
\expandafter\def\csname val@rev.desc.d7.sspro.tau-T-by-type.p2.dual-d2.text.tau.ci95.0\endcsname{0.024}
\expandafter\def\csname cnt@rev.desc.d7.sspro.tau-T-by-type.p2.dual-d2.text.tau.ci95.0\endcsname{0}
\expandafter\def\csname pct@rev.desc.d7.sspro.tau-T-by-type.p2.dual-d2.text.tau.ci95.0\endcsname{2.4}
\expandafter\def\csname ppt@rev.desc.d7.sspro.tau-T-by-type.p2.dual-d2.text.tau.ci95.0\endcsname{$+$2.4}
\expandafter\def\csname val@rev.desc.d7.sspro.tau-T-by-type.p2.dual-d2.text.tau.ci95.1\endcsname{0.190}
\expandafter\def\csname cnt@rev.desc.d7.sspro.tau-T-by-type.p2.dual-d2.text.tau.ci95.1\endcsname{0}
\expandafter\def\csname pct@rev.desc.d7.sspro.tau-T-by-type.p2.dual-d2.text.tau.ci95.1\endcsname{19.0}
\expandafter\def\csname ppt@rev.desc.d7.sspro.tau-T-by-type.p2.dual-d2.text.tau.ci95.1\endcsname{$+$19.0}
\expandafter\def\csname val@rev.desc.d7.sspro.tau-T-by-type.p2.dual-d2.text.tau.mean\endcsname{0.100}
\expandafter\def\csname cnt@rev.desc.d7.sspro.tau-T-by-type.p2.dual-d2.text.tau.mean\endcsname{0}
\expandafter\def\csname pct@rev.desc.d7.sspro.tau-T-by-type.p2.dual-d2.text.tau.mean\endcsname{10.0}
\expandafter\def\csname ppt@rev.desc.d7.sspro.tau-T-by-type.p2.dual-d2.text.tau.mean\endcsname{$+$10.0}
\expandafter\def\csname val@rev.desc.d7.sspro.tau-T-by-type.p2.dual-d2.text.tau.n-apps\endcsname{12}
\expandafter\def\csname cnt@rev.desc.d7.sspro.tau-T-by-type.p2.dual-d2.text.tau.n-apps\endcsname{12}
\expandafter\def\csname pct@rev.desc.d7.sspro.tau-T-by-type.p2.dual-d2.text.tau.n-apps\endcsname{1200.0}
\expandafter\def\csname ppt@rev.desc.d7.sspro.tau-T-by-type.p2.dual-d2.text.tau.n-apps\endcsname{$+$1200.0}
\expandafter\def\csname val@rev.desc.d7.sspro.tau-T-by-type.p2.dual-d2.text.tau.tost1.ci90-inside\endcsname{0}
\expandafter\def\csname cnt@rev.desc.d7.sspro.tau-T-by-type.p2.dual-d2.text.tau.tost1.ci90-inside\endcsname{0}
\expandafter\def\csname pct@rev.desc.d7.sspro.tau-T-by-type.p2.dual-d2.text.tau.tost1.ci90-inside\endcsname{0.0}
\expandafter\def\csname ppt@rev.desc.d7.sspro.tau-T-by-type.p2.dual-d2.text.tau.tost1.ci90-inside\endcsname{0.0}
\expandafter\def\csname val@rev.desc.d7.sspro.tau-T-by-type.p2.dual-d2.text.tau.tost1.equivalent\endcsname{0}
\expandafter\def\csname cnt@rev.desc.d7.sspro.tau-T-by-type.p2.dual-d2.text.tau.tost1.equivalent\endcsname{0}
\expandafter\def\csname pct@rev.desc.d7.sspro.tau-T-by-type.p2.dual-d2.text.tau.tost1.equivalent\endcsname{0.0}
\expandafter\def\csname ppt@rev.desc.d7.sspro.tau-T-by-type.p2.dual-d2.text.tau.tost1.equivalent\endcsname{0.0}
\expandafter\def\csname val@rev.desc.d7.sspro.tau-T-by-type.p2.dual-d2.text.tau.tost1.p\endcsname{0.991}
\expandafter\def\csname cnt@rev.desc.d7.sspro.tau-T-by-type.p2.dual-d2.text.tau.tost1.p\endcsname{1}
\expandafter\def\csname pct@rev.desc.d7.sspro.tau-T-by-type.p2.dual-d2.text.tau.tost1.p\endcsname{99.1}
\expandafter\def\csname ppt@rev.desc.d7.sspro.tau-T-by-type.p2.dual-d2.text.tau.tost1.p\endcsname{$+$99.1}
\expandafter\def\csname val@rev.desc.d7.sspro.tau-T-by-type.p2.dual-d2.text.tau.tost5.ci90-inside\endcsname{0}
\expandafter\def\csname cnt@rev.desc.d7.sspro.tau-T-by-type.p2.dual-d2.text.tau.tost5.ci90-inside\endcsname{0}
\expandafter\def\csname pct@rev.desc.d7.sspro.tau-T-by-type.p2.dual-d2.text.tau.tost5.ci90-inside\endcsname{0.0}
\expandafter\def\csname ppt@rev.desc.d7.sspro.tau-T-by-type.p2.dual-d2.text.tau.tost5.ci90-inside\endcsname{0.0}
\expandafter\def\csname val@rev.desc.d7.sspro.tau-T-by-type.p2.dual-d2.text.tau.tost5.equivalent\endcsname{0}
\expandafter\def\csname cnt@rev.desc.d7.sspro.tau-T-by-type.p2.dual-d2.text.tau.tost5.equivalent\endcsname{0}
\expandafter\def\csname pct@rev.desc.d7.sspro.tau-T-by-type.p2.dual-d2.text.tau.tost5.equivalent\endcsname{0.0}
\expandafter\def\csname ppt@rev.desc.d7.sspro.tau-T-by-type.p2.dual-d2.text.tau.tost5.equivalent\endcsname{0.0}
\expandafter\def\csname val@rev.desc.d7.sspro.tau-T-by-type.p2.dual-d2.text.tau.tost5.p\endcsname{0.885}
\expandafter\def\csname cnt@rev.desc.d7.sspro.tau-T-by-type.p2.dual-d2.text.tau.tost5.p\endcsname{1}
\expandafter\def\csname pct@rev.desc.d7.sspro.tau-T-by-type.p2.dual-d2.text.tau.tost5.p\endcsname{88.5}
\expandafter\def\csname ppt@rev.desc.d7.sspro.tau-T-by-type.p2.dual-d2.text.tau.tost5.p\endcsname{$+$88.5}
\expandafter\def\csname val@rev.desc.d7.sspro.tau-T-by-type.p2.grid2-d3.icon-minus-text-on-apps-with-both.ci90.0\endcsname{-0.020}
\expandafter\def\csname cnt@rev.desc.d7.sspro.tau-T-by-type.p2.grid2-d3.icon-minus-text-on-apps-with-both.ci90.0\endcsname{0}
\expandafter\def\csname pct@rev.desc.d7.sspro.tau-T-by-type.p2.grid2-d3.icon-minus-text-on-apps-with-both.ci90.0\endcsname{$-$2.0}
\expandafter\def\csname ppt@rev.desc.d7.sspro.tau-T-by-type.p2.grid2-d3.icon-minus-text-on-apps-with-both.ci90.0\endcsname{$-$2.0}
\expandafter\def\csname val@rev.desc.d7.sspro.tau-T-by-type.p2.grid2-d3.icon-minus-text-on-apps-with-both.ci90.1\endcsname{0.353}
\expandafter\def\csname cnt@rev.desc.d7.sspro.tau-T-by-type.p2.grid2-d3.icon-minus-text-on-apps-with-both.ci90.1\endcsname{0}
\expandafter\def\csname pct@rev.desc.d7.sspro.tau-T-by-type.p2.grid2-d3.icon-minus-text-on-apps-with-both.ci90.1\endcsname{35.3}
\expandafter\def\csname ppt@rev.desc.d7.sspro.tau-T-by-type.p2.grid2-d3.icon-minus-text-on-apps-with-both.ci90.1\endcsname{$+$35.3}
\expandafter\def\csname val@rev.desc.d7.sspro.tau-T-by-type.p2.grid2-d3.icon-minus-text-on-apps-with-both.ci95-excludes-0\endcsname{0}
\expandafter\def\csname cnt@rev.desc.d7.sspro.tau-T-by-type.p2.grid2-d3.icon-minus-text-on-apps-with-both.ci95-excludes-0\endcsname{0}
\expandafter\def\csname pct@rev.desc.d7.sspro.tau-T-by-type.p2.grid2-d3.icon-minus-text-on-apps-with-both.ci95-excludes-0\endcsname{0.0}
\expandafter\def\csname ppt@rev.desc.d7.sspro.tau-T-by-type.p2.grid2-d3.icon-minus-text-on-apps-with-both.ci95-excludes-0\endcsname{0.0}
\expandafter\def\csname val@rev.desc.d7.sspro.tau-T-by-type.p2.grid2-d3.icon-minus-text-on-apps-with-both.ci95.0\endcsname{-0.052}
\expandafter\def\csname cnt@rev.desc.d7.sspro.tau-T-by-type.p2.grid2-d3.icon-minus-text-on-apps-with-both.ci95.0\endcsname{0}
\expandafter\def\csname pct@rev.desc.d7.sspro.tau-T-by-type.p2.grid2-d3.icon-minus-text-on-apps-with-both.ci95.0\endcsname{$-$5.2}
\expandafter\def\csname ppt@rev.desc.d7.sspro.tau-T-by-type.p2.grid2-d3.icon-minus-text-on-apps-with-both.ci95.0\endcsname{$-$5.2}
\expandafter\def\csname val@rev.desc.d7.sspro.tau-T-by-type.p2.grid2-d3.icon-minus-text-on-apps-with-both.ci95.1\endcsname{0.388}
\expandafter\def\csname cnt@rev.desc.d7.sspro.tau-T-by-type.p2.grid2-d3.icon-minus-text-on-apps-with-both.ci95.1\endcsname{0}
\expandafter\def\csname pct@rev.desc.d7.sspro.tau-T-by-type.p2.grid2-d3.icon-minus-text-on-apps-with-both.ci95.1\endcsname{38.8}
\expandafter\def\csname ppt@rev.desc.d7.sspro.tau-T-by-type.p2.grid2-d3.icon-minus-text-on-apps-with-both.ci95.1\endcsname{$+$38.8}
\expandafter\def\csname val@rev.desc.d7.sspro.tau-T-by-type.p2.grid2-d3.icon-minus-text-on-apps-with-both.mean\endcsname{0.155}
\expandafter\def\csname cnt@rev.desc.d7.sspro.tau-T-by-type.p2.grid2-d3.icon-minus-text-on-apps-with-both.mean\endcsname{0}
\expandafter\def\csname pct@rev.desc.d7.sspro.tau-T-by-type.p2.grid2-d3.icon-minus-text-on-apps-with-both.mean\endcsname{15.5}
\expandafter\def\csname ppt@rev.desc.d7.sspro.tau-T-by-type.p2.grid2-d3.icon-minus-text-on-apps-with-both.mean\endcsname{$+$15.5}
\expandafter\def\csname val@rev.desc.d7.sspro.tau-T-by-type.p2.grid2-d3.icon-minus-text-on-apps-with-both.n-apps\endcsname{10}
\expandafter\def\csname cnt@rev.desc.d7.sspro.tau-T-by-type.p2.grid2-d3.icon-minus-text-on-apps-with-both.n-apps\endcsname{10}
\expandafter\def\csname pct@rev.desc.d7.sspro.tau-T-by-type.p2.grid2-d3.icon-minus-text-on-apps-with-both.n-apps\endcsname{1000.0}
\expandafter\def\csname ppt@rev.desc.d7.sspro.tau-T-by-type.p2.grid2-d3.icon-minus-text-on-apps-with-both.n-apps\endcsname{$+$1000.0}
\expandafter\def\csname val@rev.desc.d7.sspro.tau-T-by-type.p2.grid2-d3.icon-minus-text-on-apps-with-both.tost1.ci90-inside\endcsname{0}
\expandafter\def\csname cnt@rev.desc.d7.sspro.tau-T-by-type.p2.grid2-d3.icon-minus-text-on-apps-with-both.tost1.ci90-inside\endcsname{0}
\expandafter\def\csname pct@rev.desc.d7.sspro.tau-T-by-type.p2.grid2-d3.icon-minus-text-on-apps-with-both.tost1.ci90-inside\endcsname{0.0}
\expandafter\def\csname ppt@rev.desc.d7.sspro.tau-T-by-type.p2.grid2-d3.icon-minus-text-on-apps-with-both.tost1.ci90-inside\endcsname{0.0}
\expandafter\def\csname val@rev.desc.d7.sspro.tau-T-by-type.p2.grid2-d3.icon-minus-text-on-apps-with-both.tost1.equivalent\endcsname{0}
\expandafter\def\csname cnt@rev.desc.d7.sspro.tau-T-by-type.p2.grid2-d3.icon-minus-text-on-apps-with-both.tost1.equivalent\endcsname{0}
\expandafter\def\csname pct@rev.desc.d7.sspro.tau-T-by-type.p2.grid2-d3.icon-minus-text-on-apps-with-both.tost1.equivalent\endcsname{0.0}
\expandafter\def\csname ppt@rev.desc.d7.sspro.tau-T-by-type.p2.grid2-d3.icon-minus-text-on-apps-with-both.tost1.equivalent\endcsname{0.0}
\expandafter\def\csname val@rev.desc.d7.sspro.tau-T-by-type.p2.grid2-d3.icon-minus-text-on-apps-with-both.tost1.p\endcsname{0.903}
\expandafter\def\csname cnt@rev.desc.d7.sspro.tau-T-by-type.p2.grid2-d3.icon-minus-text-on-apps-with-both.tost1.p\endcsname{1}
\expandafter\def\csname pct@rev.desc.d7.sspro.tau-T-by-type.p2.grid2-d3.icon-minus-text-on-apps-with-both.tost1.p\endcsname{90.3}
\expandafter\def\csname ppt@rev.desc.d7.sspro.tau-T-by-type.p2.grid2-d3.icon-minus-text-on-apps-with-both.tost1.p\endcsname{$+$90.3}
\expandafter\def\csname val@rev.desc.d7.sspro.tau-T-by-type.p2.grid2-d3.icon-minus-text-on-apps-with-both.tost5.ci90-inside\endcsname{0}
\expandafter\def\csname cnt@rev.desc.d7.sspro.tau-T-by-type.p2.grid2-d3.icon-minus-text-on-apps-with-both.tost5.ci90-inside\endcsname{0}
\expandafter\def\csname pct@rev.desc.d7.sspro.tau-T-by-type.p2.grid2-d3.icon-minus-text-on-apps-with-both.tost5.ci90-inside\endcsname{0.0}
\expandafter\def\csname ppt@rev.desc.d7.sspro.tau-T-by-type.p2.grid2-d3.icon-minus-text-on-apps-with-both.tost5.ci90-inside\endcsname{0.0}
\expandafter\def\csname val@rev.desc.d7.sspro.tau-T-by-type.p2.grid2-d3.icon-minus-text-on-apps-with-both.tost5.equivalent\endcsname{0}
\expandafter\def\csname cnt@rev.desc.d7.sspro.tau-T-by-type.p2.grid2-d3.icon-minus-text-on-apps-with-both.tost5.equivalent\endcsname{0}
\expandafter\def\csname pct@rev.desc.d7.sspro.tau-T-by-type.p2.grid2-d3.icon-minus-text-on-apps-with-both.tost5.equivalent\endcsname{0.0}
\expandafter\def\csname ppt@rev.desc.d7.sspro.tau-T-by-type.p2.grid2-d3.icon-minus-text-on-apps-with-both.tost5.equivalent\endcsname{0.0}
\expandafter\def\csname val@rev.desc.d7.sspro.tau-T-by-type.p2.grid2-d3.icon-minus-text-on-apps-with-both.tost5.p\endcsname{0.826}
\expandafter\def\csname cnt@rev.desc.d7.sspro.tau-T-by-type.p2.grid2-d3.icon-minus-text-on-apps-with-both.tost5.p\endcsname{1}
\expandafter\def\csname pct@rev.desc.d7.sspro.tau-T-by-type.p2.grid2-d3.icon-minus-text-on-apps-with-both.tost5.p\endcsname{82.6}
\expandafter\def\csname ppt@rev.desc.d7.sspro.tau-T-by-type.p2.grid2-d3.icon-minus-text-on-apps-with-both.tost5.p\endcsname{$+$82.6}
\expandafter\def\csname val@rev.desc.d7.sspro.tau-T-by-type.p2.grid2-d3.icon.acc-F\endcsname{0.714}
\expandafter\def\csname cnt@rev.desc.d7.sspro.tau-T-by-type.p2.grid2-d3.icon.acc-F\endcsname{1}
\expandafter\def\csname pct@rev.desc.d7.sspro.tau-T-by-type.p2.grid2-d3.icon.acc-F\endcsname{71.4}
\expandafter\def\csname ppt@rev.desc.d7.sspro.tau-T-by-type.p2.grid2-d3.icon.acc-F\endcsname{$+$71.4}
\expandafter\def\csname val@rev.desc.d7.sspro.tau-T-by-type.p2.grid2-d3.icon.acc-direct\endcsname{0.571}
\expandafter\def\csname cnt@rev.desc.d7.sspro.tau-T-by-type.p2.grid2-d3.icon.acc-direct\endcsname{1}
\expandafter\def\csname pct@rev.desc.d7.sspro.tau-T-by-type.p2.grid2-d3.icon.acc-direct\endcsname{57.1}
\expandafter\def\csname ppt@rev.desc.d7.sspro.tau-T-by-type.p2.grid2-d3.icon.acc-direct\endcsname{$+$57.1}
\expandafter\def\csname val@rev.desc.d7.sspro.tau-T-by-type.p2.grid2-d3.icon.n-apps\endcsname{10}
\expandafter\def\csname cnt@rev.desc.d7.sspro.tau-T-by-type.p2.grid2-d3.icon.n-apps\endcsname{10}
\expandafter\def\csname pct@rev.desc.d7.sspro.tau-T-by-type.p2.grid2-d3.icon.n-apps\endcsname{1000.0}
\expandafter\def\csname ppt@rev.desc.d7.sspro.tau-T-by-type.p2.grid2-d3.icon.n-apps\endcsname{$+$1000.0}
\expandafter\def\csname val@rev.desc.d7.sspro.tau-T-by-type.p2.grid2-d3.icon.n-tasks\endcsname{28}
\expandafter\def\csname cnt@rev.desc.d7.sspro.tau-T-by-type.p2.grid2-d3.icon.n-tasks\endcsname{28}
\expandafter\def\csname pct@rev.desc.d7.sspro.tau-T-by-type.p2.grid2-d3.icon.n-tasks\endcsname{2800.0}
\expandafter\def\csname ppt@rev.desc.d7.sspro.tau-T-by-type.p2.grid2-d3.icon.n-tasks\endcsname{$+$2800.0}
\expandafter\def\csname val@rev.desc.d7.sspro.tau-T-by-type.p2.grid2-d3.icon.tau.ci90.0\endcsname{0.063}
\expandafter\def\csname cnt@rev.desc.d7.sspro.tau-T-by-type.p2.grid2-d3.icon.tau.ci90.0\endcsname{0}
\expandafter\def\csname pct@rev.desc.d7.sspro.tau-T-by-type.p2.grid2-d3.icon.tau.ci90.0\endcsname{6.3}
\expandafter\def\csname ppt@rev.desc.d7.sspro.tau-T-by-type.p2.grid2-d3.icon.tau.ci90.0\endcsname{$+$6.3}
\expandafter\def\csname val@rev.desc.d7.sspro.tau-T-by-type.p2.grid2-d3.icon.tau.ci90.1\endcsname{0.400}
\expandafter\def\csname cnt@rev.desc.d7.sspro.tau-T-by-type.p2.grid2-d3.icon.tau.ci90.1\endcsname{0}
\expandafter\def\csname pct@rev.desc.d7.sspro.tau-T-by-type.p2.grid2-d3.icon.tau.ci90.1\endcsname{40.0}
\expandafter\def\csname ppt@rev.desc.d7.sspro.tau-T-by-type.p2.grid2-d3.icon.tau.ci90.1\endcsname{$+$40.0}
\expandafter\def\csname val@rev.desc.d7.sspro.tau-T-by-type.p2.grid2-d3.icon.tau.ci95-excludes-0\endcsname{1}
\expandafter\def\csname cnt@rev.desc.d7.sspro.tau-T-by-type.p2.grid2-d3.icon.tau.ci95-excludes-0\endcsname{1}
\expandafter\def\csname pct@rev.desc.d7.sspro.tau-T-by-type.p2.grid2-d3.icon.tau.ci95-excludes-0\endcsname{100.0}
\expandafter\def\csname ppt@rev.desc.d7.sspro.tau-T-by-type.p2.grid2-d3.icon.tau.ci95-excludes-0\endcsname{$+$100.0}
\expandafter\def\csname val@rev.desc.d7.sspro.tau-T-by-type.p2.grid2-d3.icon.tau.ci95.0\endcsname{0.040}
\expandafter\def\csname cnt@rev.desc.d7.sspro.tau-T-by-type.p2.grid2-d3.icon.tau.ci95.0\endcsname{0}
\expandafter\def\csname pct@rev.desc.d7.sspro.tau-T-by-type.p2.grid2-d3.icon.tau.ci95.0\endcsname{4.0}
\expandafter\def\csname ppt@rev.desc.d7.sspro.tau-T-by-type.p2.grid2-d3.icon.tau.ci95.0\endcsname{$+$4.0}
\expandafter\def\csname val@rev.desc.d7.sspro.tau-T-by-type.p2.grid2-d3.icon.tau.ci95.1\endcsname{0.433}
\expandafter\def\csname cnt@rev.desc.d7.sspro.tau-T-by-type.p2.grid2-d3.icon.tau.ci95.1\endcsname{0}
\expandafter\def\csname pct@rev.desc.d7.sspro.tau-T-by-type.p2.grid2-d3.icon.tau.ci95.1\endcsname{43.3}
\expandafter\def\csname ppt@rev.desc.d7.sspro.tau-T-by-type.p2.grid2-d3.icon.tau.ci95.1\endcsname{$+$43.3}
\expandafter\def\csname val@rev.desc.d7.sspro.tau-T-by-type.p2.grid2-d3.icon.tau.mean\endcsname{0.217}
\expandafter\def\csname cnt@rev.desc.d7.sspro.tau-T-by-type.p2.grid2-d3.icon.tau.mean\endcsname{0}
\expandafter\def\csname pct@rev.desc.d7.sspro.tau-T-by-type.p2.grid2-d3.icon.tau.mean\endcsname{21.7}
\expandafter\def\csname ppt@rev.desc.d7.sspro.tau-T-by-type.p2.grid2-d3.icon.tau.mean\endcsname{$+$21.7}
\expandafter\def\csname val@rev.desc.d7.sspro.tau-T-by-type.p2.grid2-d3.icon.tau.n-apps\endcsname{10}
\expandafter\def\csname cnt@rev.desc.d7.sspro.tau-T-by-type.p2.grid2-d3.icon.tau.n-apps\endcsname{10}
\expandafter\def\csname pct@rev.desc.d7.sspro.tau-T-by-type.p2.grid2-d3.icon.tau.n-apps\endcsname{1000.0}
\expandafter\def\csname ppt@rev.desc.d7.sspro.tau-T-by-type.p2.grid2-d3.icon.tau.n-apps\endcsname{$+$1000.0}
\expandafter\def\csname val@rev.desc.d7.sspro.tau-T-by-type.p2.grid2-d3.icon.tau.tost1.ci90-inside\endcsname{0}
\expandafter\def\csname cnt@rev.desc.d7.sspro.tau-T-by-type.p2.grid2-d3.icon.tau.tost1.ci90-inside\endcsname{0}
\expandafter\def\csname pct@rev.desc.d7.sspro.tau-T-by-type.p2.grid2-d3.icon.tau.tost1.ci90-inside\endcsname{0.0}
\expandafter\def\csname ppt@rev.desc.d7.sspro.tau-T-by-type.p2.grid2-d3.icon.tau.tost1.ci90-inside\endcsname{0.0}
\expandafter\def\csname val@rev.desc.d7.sspro.tau-T-by-type.p2.grid2-d3.icon.tau.tost1.equivalent\endcsname{0}
\expandafter\def\csname cnt@rev.desc.d7.sspro.tau-T-by-type.p2.grid2-d3.icon.tau.tost1.equivalent\endcsname{0}
\expandafter\def\csname pct@rev.desc.d7.sspro.tau-T-by-type.p2.grid2-d3.icon.tau.tost1.equivalent\endcsname{0.0}
\expandafter\def\csname ppt@rev.desc.d7.sspro.tau-T-by-type.p2.grid2-d3.icon.tau.tost1.equivalent\endcsname{0.0}
\expandafter\def\csname val@rev.desc.d7.sspro.tau-T-by-type.p2.grid2-d3.icon.tau.tost1.p\endcsname{0.992}
\expandafter\def\csname cnt@rev.desc.d7.sspro.tau-T-by-type.p2.grid2-d3.icon.tau.tost1.p\endcsname{1}
\expandafter\def\csname pct@rev.desc.d7.sspro.tau-T-by-type.p2.grid2-d3.icon.tau.tost1.p\endcsname{99.2}
\expandafter\def\csname ppt@rev.desc.d7.sspro.tau-T-by-type.p2.grid2-d3.icon.tau.tost1.p\endcsname{$+$99.2}
\expandafter\def\csname val@rev.desc.d7.sspro.tau-T-by-type.p2.grid2-d3.icon.tau.tost5.ci90-inside\endcsname{0}
\expandafter\def\csname cnt@rev.desc.d7.sspro.tau-T-by-type.p2.grid2-d3.icon.tau.tost5.ci90-inside\endcsname{0}
\expandafter\def\csname pct@rev.desc.d7.sspro.tau-T-by-type.p2.grid2-d3.icon.tau.tost5.ci90-inside\endcsname{0.0}
\expandafter\def\csname ppt@rev.desc.d7.sspro.tau-T-by-type.p2.grid2-d3.icon.tau.tost5.ci90-inside\endcsname{0.0}
\expandafter\def\csname val@rev.desc.d7.sspro.tau-T-by-type.p2.grid2-d3.icon.tau.tost5.equivalent\endcsname{0}
\expandafter\def\csname cnt@rev.desc.d7.sspro.tau-T-by-type.p2.grid2-d3.icon.tau.tost5.equivalent\endcsname{0}
\expandafter\def\csname pct@rev.desc.d7.sspro.tau-T-by-type.p2.grid2-d3.icon.tau.tost5.equivalent\endcsname{0.0}
\expandafter\def\csname ppt@rev.desc.d7.sspro.tau-T-by-type.p2.grid2-d3.icon.tau.tost5.equivalent\endcsname{0.0}
\expandafter\def\csname val@rev.desc.d7.sspro.tau-T-by-type.p2.grid2-d3.icon.tau.tost5.p\endcsname{0.963}
\expandafter\def\csname cnt@rev.desc.d7.sspro.tau-T-by-type.p2.grid2-d3.icon.tau.tost5.p\endcsname{1}
\expandafter\def\csname pct@rev.desc.d7.sspro.tau-T-by-type.p2.grid2-d3.icon.tau.tost5.p\endcsname{96.3}
\expandafter\def\csname ppt@rev.desc.d7.sspro.tau-T-by-type.p2.grid2-d3.icon.tau.tost5.p\endcsname{$+$96.3}
\expandafter\def\csname val@rev.desc.d7.sspro.tau-T-by-type.p2.grid2-d3.text.acc-F\endcsname{0.985}
\expandafter\def\csname cnt@rev.desc.d7.sspro.tau-T-by-type.p2.grid2-d3.text.acc-F\endcsname{1}
\expandafter\def\csname pct@rev.desc.d7.sspro.tau-T-by-type.p2.grid2-d3.text.acc-F\endcsname{98.5}
\expandafter\def\csname ppt@rev.desc.d7.sspro.tau-T-by-type.p2.grid2-d3.text.acc-F\endcsname{$+$98.5}
\expandafter\def\csname val@rev.desc.d7.sspro.tau-T-by-type.p2.grid2-d3.text.acc-direct\endcsname{0.838}
\expandafter\def\csname cnt@rev.desc.d7.sspro.tau-T-by-type.p2.grid2-d3.text.acc-direct\endcsname{1}
\expandafter\def\csname pct@rev.desc.d7.sspro.tau-T-by-type.p2.grid2-d3.text.acc-direct\endcsname{83.8}
\expandafter\def\csname ppt@rev.desc.d7.sspro.tau-T-by-type.p2.grid2-d3.text.acc-direct\endcsname{$+$83.8}
\expandafter\def\csname val@rev.desc.d7.sspro.tau-T-by-type.p2.grid2-d3.text.n-apps\endcsname{12}
\expandafter\def\csname cnt@rev.desc.d7.sspro.tau-T-by-type.p2.grid2-d3.text.n-apps\endcsname{12}
\expandafter\def\csname pct@rev.desc.d7.sspro.tau-T-by-type.p2.grid2-d3.text.n-apps\endcsname{1200.0}
\expandafter\def\csname ppt@rev.desc.d7.sspro.tau-T-by-type.p2.grid2-d3.text.n-apps\endcsname{$+$1200.0}
\expandafter\def\csname val@rev.desc.d7.sspro.tau-T-by-type.p2.grid2-d3.text.n-tasks\endcsname{68}
\expandafter\def\csname cnt@rev.desc.d7.sspro.tau-T-by-type.p2.grid2-d3.text.n-tasks\endcsname{68}
\expandafter\def\csname pct@rev.desc.d7.sspro.tau-T-by-type.p2.grid2-d3.text.n-tasks\endcsname{6800.0}
\expandafter\def\csname ppt@rev.desc.d7.sspro.tau-T-by-type.p2.grid2-d3.text.n-tasks\endcsname{$+$6800.0}
\expandafter\def\csname val@rev.desc.d7.sspro.tau-T-by-type.p2.grid2-d3.text.tau.ci90.0\endcsname{0.042}
\expandafter\def\csname cnt@rev.desc.d7.sspro.tau-T-by-type.p2.grid2-d3.text.tau.ci90.0\endcsname{0}
\expandafter\def\csname pct@rev.desc.d7.sspro.tau-T-by-type.p2.grid2-d3.text.tau.ci90.0\endcsname{4.2}
\expandafter\def\csname ppt@rev.desc.d7.sspro.tau-T-by-type.p2.grid2-d3.text.tau.ci90.0\endcsname{$+$4.2}
\expandafter\def\csname val@rev.desc.d7.sspro.tau-T-by-type.p2.grid2-d3.text.tau.ci90.1\endcsname{0.193}
\expandafter\def\csname cnt@rev.desc.d7.sspro.tau-T-by-type.p2.grid2-d3.text.tau.ci90.1\endcsname{0}
\expandafter\def\csname pct@rev.desc.d7.sspro.tau-T-by-type.p2.grid2-d3.text.tau.ci90.1\endcsname{19.3}
\expandafter\def\csname ppt@rev.desc.d7.sspro.tau-T-by-type.p2.grid2-d3.text.tau.ci90.1\endcsname{$+$19.3}
\expandafter\def\csname val@rev.desc.d7.sspro.tau-T-by-type.p2.grid2-d3.text.tau.ci95-excludes-0\endcsname{1}
\expandafter\def\csname cnt@rev.desc.d7.sspro.tau-T-by-type.p2.grid2-d3.text.tau.ci95-excludes-0\endcsname{1}
\expandafter\def\csname pct@rev.desc.d7.sspro.tau-T-by-type.p2.grid2-d3.text.tau.ci95-excludes-0\endcsname{100.0}
\expandafter\def\csname ppt@rev.desc.d7.sspro.tau-T-by-type.p2.grid2-d3.text.tau.ci95-excludes-0\endcsname{$+$100.0}
\expandafter\def\csname val@rev.desc.d7.sspro.tau-T-by-type.p2.grid2-d3.text.tau.ci95.0\endcsname{0.035}
\expandafter\def\csname cnt@rev.desc.d7.sspro.tau-T-by-type.p2.grid2-d3.text.tau.ci95.0\endcsname{0}
\expandafter\def\csname pct@rev.desc.d7.sspro.tau-T-by-type.p2.grid2-d3.text.tau.ci95.0\endcsname{3.5}
\expandafter\def\csname ppt@rev.desc.d7.sspro.tau-T-by-type.p2.grid2-d3.text.tau.ci95.0\endcsname{$+$3.5}
\expandafter\def\csname val@rev.desc.d7.sspro.tau-T-by-type.p2.grid2-d3.text.tau.ci95.1\endcsname{0.208}
\expandafter\def\csname cnt@rev.desc.d7.sspro.tau-T-by-type.p2.grid2-d3.text.tau.ci95.1\endcsname{0}
\expandafter\def\csname pct@rev.desc.d7.sspro.tau-T-by-type.p2.grid2-d3.text.tau.ci95.1\endcsname{20.8}
\expandafter\def\csname ppt@rev.desc.d7.sspro.tau-T-by-type.p2.grid2-d3.text.tau.ci95.1\endcsname{$+$20.8}
\expandafter\def\csname val@rev.desc.d7.sspro.tau-T-by-type.p2.grid2-d3.text.tau.mean\endcsname{0.114}
\expandafter\def\csname cnt@rev.desc.d7.sspro.tau-T-by-type.p2.grid2-d3.text.tau.mean\endcsname{0}
\expandafter\def\csname pct@rev.desc.d7.sspro.tau-T-by-type.p2.grid2-d3.text.tau.mean\endcsname{11.4}
\expandafter\def\csname ppt@rev.desc.d7.sspro.tau-T-by-type.p2.grid2-d3.text.tau.mean\endcsname{$+$11.4}
\expandafter\def\csname val@rev.desc.d7.sspro.tau-T-by-type.p2.grid2-d3.text.tau.n-apps\endcsname{12}
\expandafter\def\csname cnt@rev.desc.d7.sspro.tau-T-by-type.p2.grid2-d3.text.tau.n-apps\endcsname{12}
\expandafter\def\csname pct@rev.desc.d7.sspro.tau-T-by-type.p2.grid2-d3.text.tau.n-apps\endcsname{1200.0}
\expandafter\def\csname ppt@rev.desc.d7.sspro.tau-T-by-type.p2.grid2-d3.text.tau.n-apps\endcsname{$+$1200.0}
\expandafter\def\csname val@rev.desc.d7.sspro.tau-T-by-type.p2.grid2-d3.text.tau.tost1.ci90-inside\endcsname{0}
\expandafter\def\csname cnt@rev.desc.d7.sspro.tau-T-by-type.p2.grid2-d3.text.tau.tost1.ci90-inside\endcsname{0}
\expandafter\def\csname pct@rev.desc.d7.sspro.tau-T-by-type.p2.grid2-d3.text.tau.tost1.ci90-inside\endcsname{0.0}
\expandafter\def\csname ppt@rev.desc.d7.sspro.tau-T-by-type.p2.grid2-d3.text.tau.tost1.ci90-inside\endcsname{0.0}
\expandafter\def\csname val@rev.desc.d7.sspro.tau-T-by-type.p2.grid2-d3.text.tau.tost1.equivalent\endcsname{0}
\expandafter\def\csname cnt@rev.desc.d7.sspro.tau-T-by-type.p2.grid2-d3.text.tau.tost1.equivalent\endcsname{0}
\expandafter\def\csname pct@rev.desc.d7.sspro.tau-T-by-type.p2.grid2-d3.text.tau.tost1.equivalent\endcsname{0.0}
\expandafter\def\csname ppt@rev.desc.d7.sspro.tau-T-by-type.p2.grid2-d3.text.tau.tost1.equivalent\endcsname{0.0}
\expandafter\def\csname val@rev.desc.d7.sspro.tau-T-by-type.p2.grid2-d3.text.tau.tost1.p\endcsname{0.999}
\expandafter\def\csname cnt@rev.desc.d7.sspro.tau-T-by-type.p2.grid2-d3.text.tau.tost1.p\endcsname{1}
\expandafter\def\csname pct@rev.desc.d7.sspro.tau-T-by-type.p2.grid2-d3.text.tau.tost1.p\endcsname{99.9}
\expandafter\def\csname ppt@rev.desc.d7.sspro.tau-T-by-type.p2.grid2-d3.text.tau.tost1.p\endcsname{$+$99.9}
\expandafter\def\csname val@rev.desc.d7.sspro.tau-T-by-type.p2.grid2-d3.text.tau.tost5.ci90-inside\endcsname{0}
\expandafter\def\csname cnt@rev.desc.d7.sspro.tau-T-by-type.p2.grid2-d3.text.tau.tost5.ci90-inside\endcsname{0}
\expandafter\def\csname pct@rev.desc.d7.sspro.tau-T-by-type.p2.grid2-d3.text.tau.tost5.ci90-inside\endcsname{0.0}
\expandafter\def\csname ppt@rev.desc.d7.sspro.tau-T-by-type.p2.grid2-d3.text.tau.tost5.ci90-inside\endcsname{0.0}
\expandafter\def\csname val@rev.desc.d7.sspro.tau-T-by-type.p2.grid2-d3.text.tau.tost5.equivalent\endcsname{0}
\expandafter\def\csname cnt@rev.desc.d7.sspro.tau-T-by-type.p2.grid2-d3.text.tau.tost5.equivalent\endcsname{0}
\expandafter\def\csname pct@rev.desc.d7.sspro.tau-T-by-type.p2.grid2-d3.text.tau.tost5.equivalent\endcsname{0.0}
\expandafter\def\csname ppt@rev.desc.d7.sspro.tau-T-by-type.p2.grid2-d3.text.tau.tost5.equivalent\endcsname{0.0}
\expandafter\def\csname val@rev.desc.d7.sspro.tau-T-by-type.p2.grid2-d3.text.tau.tost5.p\endcsname{0.929}
\expandafter\def\csname cnt@rev.desc.d7.sspro.tau-T-by-type.p2.grid2-d3.text.tau.tost5.p\endcsname{1}
\expandafter\def\csname pct@rev.desc.d7.sspro.tau-T-by-type.p2.grid2-d3.text.tau.tost5.p\endcsname{92.9}
\expandafter\def\csname ppt@rev.desc.d7.sspro.tau-T-by-type.p2.grid2-d3.text.tau.tost5.p\endcsname{$+$92.9}
\expandafter\def\csname val@rev.desc.d7.sspro.tau-T-by-type.p2.zoomclick-d3.icon-minus-text-on-apps-with-both.ci90.0\endcsname{0.055}
\expandafter\def\csname cnt@rev.desc.d7.sspro.tau-T-by-type.p2.zoomclick-d3.icon-minus-text-on-apps-with-both.ci90.0\endcsname{0}
\expandafter\def\csname pct@rev.desc.d7.sspro.tau-T-by-type.p2.zoomclick-d3.icon-minus-text-on-apps-with-both.ci90.0\endcsname{5.5}
\expandafter\def\csname ppt@rev.desc.d7.sspro.tau-T-by-type.p2.zoomclick-d3.icon-minus-text-on-apps-with-both.ci90.0\endcsname{$+$5.5}
\expandafter\def\csname val@rev.desc.d7.sspro.tau-T-by-type.p2.zoomclick-d3.icon-minus-text-on-apps-with-both.ci90.1\endcsname{0.441}
\expandafter\def\csname cnt@rev.desc.d7.sspro.tau-T-by-type.p2.zoomclick-d3.icon-minus-text-on-apps-with-both.ci90.1\endcsname{0}
\expandafter\def\csname pct@rev.desc.d7.sspro.tau-T-by-type.p2.zoomclick-d3.icon-minus-text-on-apps-with-both.ci90.1\endcsname{44.1}
\expandafter\def\csname ppt@rev.desc.d7.sspro.tau-T-by-type.p2.zoomclick-d3.icon-minus-text-on-apps-with-both.ci90.1\endcsname{$+$44.1}
\expandafter\def\csname val@rev.desc.d7.sspro.tau-T-by-type.p2.zoomclick-d3.icon-minus-text-on-apps-with-both.ci95-excludes-0\endcsname{1}
\expandafter\def\csname cnt@rev.desc.d7.sspro.tau-T-by-type.p2.zoomclick-d3.icon-minus-text-on-apps-with-both.ci95-excludes-0\endcsname{1}
\expandafter\def\csname pct@rev.desc.d7.sspro.tau-T-by-type.p2.zoomclick-d3.icon-minus-text-on-apps-with-both.ci95-excludes-0\endcsname{100.0}
\expandafter\def\csname ppt@rev.desc.d7.sspro.tau-T-by-type.p2.zoomclick-d3.icon-minus-text-on-apps-with-both.ci95-excludes-0\endcsname{$+$100.0}
\expandafter\def\csname val@rev.desc.d7.sspro.tau-T-by-type.p2.zoomclick-d3.icon-minus-text-on-apps-with-both.ci95.0\endcsname{0.027}
\expandafter\def\csname cnt@rev.desc.d7.sspro.tau-T-by-type.p2.zoomclick-d3.icon-minus-text-on-apps-with-both.ci95.0\endcsname{0}
\expandafter\def\csname pct@rev.desc.d7.sspro.tau-T-by-type.p2.zoomclick-d3.icon-minus-text-on-apps-with-both.ci95.0\endcsname{2.7}
\expandafter\def\csname ppt@rev.desc.d7.sspro.tau-T-by-type.p2.zoomclick-d3.icon-minus-text-on-apps-with-both.ci95.0\endcsname{$+$2.7}
\expandafter\def\csname val@rev.desc.d7.sspro.tau-T-by-type.p2.zoomclick-d3.icon-minus-text-on-apps-with-both.ci95.1\endcsname{0.480}
\expandafter\def\csname cnt@rev.desc.d7.sspro.tau-T-by-type.p2.zoomclick-d3.icon-minus-text-on-apps-with-both.ci95.1\endcsname{0}
\expandafter\def\csname pct@rev.desc.d7.sspro.tau-T-by-type.p2.zoomclick-d3.icon-minus-text-on-apps-with-both.ci95.1\endcsname{48.0}
\expandafter\def\csname ppt@rev.desc.d7.sspro.tau-T-by-type.p2.zoomclick-d3.icon-minus-text-on-apps-with-both.ci95.1\endcsname{$+$48.0}
\expandafter\def\csname val@rev.desc.d7.sspro.tau-T-by-type.p2.zoomclick-d3.icon-minus-text-on-apps-with-both.mean\endcsname{0.245}
\expandafter\def\csname cnt@rev.desc.d7.sspro.tau-T-by-type.p2.zoomclick-d3.icon-minus-text-on-apps-with-both.mean\endcsname{0}
\expandafter\def\csname pct@rev.desc.d7.sspro.tau-T-by-type.p2.zoomclick-d3.icon-minus-text-on-apps-with-both.mean\endcsname{24.5}
\expandafter\def\csname ppt@rev.desc.d7.sspro.tau-T-by-type.p2.zoomclick-d3.icon-minus-text-on-apps-with-both.mean\endcsname{$+$24.5}
\expandafter\def\csname val@rev.desc.d7.sspro.tau-T-by-type.p2.zoomclick-d3.icon-minus-text-on-apps-with-both.n-apps\endcsname{10}
\expandafter\def\csname cnt@rev.desc.d7.sspro.tau-T-by-type.p2.zoomclick-d3.icon-minus-text-on-apps-with-both.n-apps\endcsname{10}
\expandafter\def\csname pct@rev.desc.d7.sspro.tau-T-by-type.p2.zoomclick-d3.icon-minus-text-on-apps-with-both.n-apps\endcsname{1000.0}
\expandafter\def\csname ppt@rev.desc.d7.sspro.tau-T-by-type.p2.zoomclick-d3.icon-minus-text-on-apps-with-both.n-apps\endcsname{$+$1000.0}
\expandafter\def\csname val@rev.desc.d7.sspro.tau-T-by-type.p2.zoomclick-d3.icon-minus-text-on-apps-with-both.tost1.ci90-inside\endcsname{0}
\expandafter\def\csname cnt@rev.desc.d7.sspro.tau-T-by-type.p2.zoomclick-d3.icon-minus-text-on-apps-with-both.tost1.ci90-inside\endcsname{0}
\expandafter\def\csname pct@rev.desc.d7.sspro.tau-T-by-type.p2.zoomclick-d3.icon-minus-text-on-apps-with-both.tost1.ci90-inside\endcsname{0.0}
\expandafter\def\csname ppt@rev.desc.d7.sspro.tau-T-by-type.p2.zoomclick-d3.icon-minus-text-on-apps-with-both.tost1.ci90-inside\endcsname{0.0}
\expandafter\def\csname val@rev.desc.d7.sspro.tau-T-by-type.p2.zoomclick-d3.icon-minus-text-on-apps-with-both.tost1.equivalent\endcsname{0}
\expandafter\def\csname cnt@rev.desc.d7.sspro.tau-T-by-type.p2.zoomclick-d3.icon-minus-text-on-apps-with-both.tost1.equivalent\endcsname{0}
\expandafter\def\csname pct@rev.desc.d7.sspro.tau-T-by-type.p2.zoomclick-d3.icon-minus-text-on-apps-with-both.tost1.equivalent\endcsname{0.0}
\expandafter\def\csname ppt@rev.desc.d7.sspro.tau-T-by-type.p2.zoomclick-d3.icon-minus-text-on-apps-with-both.tost1.equivalent\endcsname{0.0}
\expandafter\def\csname val@rev.desc.d7.sspro.tau-T-by-type.p2.zoomclick-d3.icon-minus-text-on-apps-with-both.tost1.p\endcsname{0.983}
\expandafter\def\csname cnt@rev.desc.d7.sspro.tau-T-by-type.p2.zoomclick-d3.icon-minus-text-on-apps-with-both.tost1.p\endcsname{1}
\expandafter\def\csname pct@rev.desc.d7.sspro.tau-T-by-type.p2.zoomclick-d3.icon-minus-text-on-apps-with-both.tost1.p\endcsname{98.3}
\expandafter\def\csname ppt@rev.desc.d7.sspro.tau-T-by-type.p2.zoomclick-d3.icon-minus-text-on-apps-with-both.tost1.p\endcsname{$+$98.3}
\expandafter\def\csname val@rev.desc.d7.sspro.tau-T-by-type.p2.zoomclick-d3.icon-minus-text-on-apps-with-both.tost5.ci90-inside\endcsname{0}
\expandafter\def\csname cnt@rev.desc.d7.sspro.tau-T-by-type.p2.zoomclick-d3.icon-minus-text-on-apps-with-both.tost5.ci90-inside\endcsname{0}
\expandafter\def\csname pct@rev.desc.d7.sspro.tau-T-by-type.p2.zoomclick-d3.icon-minus-text-on-apps-with-both.tost5.ci90-inside\endcsname{0.0}
\expandafter\def\csname ppt@rev.desc.d7.sspro.tau-T-by-type.p2.zoomclick-d3.icon-minus-text-on-apps-with-both.tost5.ci90-inside\endcsname{0.0}
\expandafter\def\csname val@rev.desc.d7.sspro.tau-T-by-type.p2.zoomclick-d3.icon-minus-text-on-apps-with-both.tost5.equivalent\endcsname{0}
\expandafter\def\csname cnt@rev.desc.d7.sspro.tau-T-by-type.p2.zoomclick-d3.icon-minus-text-on-apps-with-both.tost5.equivalent\endcsname{0}
\expandafter\def\csname pct@rev.desc.d7.sspro.tau-T-by-type.p2.zoomclick-d3.icon-minus-text-on-apps-with-both.tost5.equivalent\endcsname{0.0}
\expandafter\def\csname ppt@rev.desc.d7.sspro.tau-T-by-type.p2.zoomclick-d3.icon-minus-text-on-apps-with-both.tost5.equivalent\endcsname{0.0}
\expandafter\def\csname val@rev.desc.d7.sspro.tau-T-by-type.p2.zoomclick-d3.icon-minus-text-on-apps-with-both.tost5.p\endcsname{0.954}
\expandafter\def\csname cnt@rev.desc.d7.sspro.tau-T-by-type.p2.zoomclick-d3.icon-minus-text-on-apps-with-both.tost5.p\endcsname{1}
\expandafter\def\csname pct@rev.desc.d7.sspro.tau-T-by-type.p2.zoomclick-d3.icon-minus-text-on-apps-with-both.tost5.p\endcsname{95.4}
\expandafter\def\csname ppt@rev.desc.d7.sspro.tau-T-by-type.p2.zoomclick-d3.icon-minus-text-on-apps-with-both.tost5.p\endcsname{$+$95.4}
\expandafter\def\csname val@rev.desc.d7.sspro.tau-T-by-type.p2.zoomclick-d3.icon.acc-F\endcsname{0.821}
\expandafter\def\csname cnt@rev.desc.d7.sspro.tau-T-by-type.p2.zoomclick-d3.icon.acc-F\endcsname{1}
\expandafter\def\csname pct@rev.desc.d7.sspro.tau-T-by-type.p2.zoomclick-d3.icon.acc-F\endcsname{82.1}
\expandafter\def\csname ppt@rev.desc.d7.sspro.tau-T-by-type.p2.zoomclick-d3.icon.acc-F\endcsname{$+$82.1}
\expandafter\def\csname val@rev.desc.d7.sspro.tau-T-by-type.p2.zoomclick-d3.icon.acc-direct\endcsname{0.571}
\expandafter\def\csname cnt@rev.desc.d7.sspro.tau-T-by-type.p2.zoomclick-d3.icon.acc-direct\endcsname{1}
\expandafter\def\csname pct@rev.desc.d7.sspro.tau-T-by-type.p2.zoomclick-d3.icon.acc-direct\endcsname{57.1}
\expandafter\def\csname ppt@rev.desc.d7.sspro.tau-T-by-type.p2.zoomclick-d3.icon.acc-direct\endcsname{$+$57.1}
\expandafter\def\csname val@rev.desc.d7.sspro.tau-T-by-type.p2.zoomclick-d3.icon.n-apps\endcsname{10}
\expandafter\def\csname cnt@rev.desc.d7.sspro.tau-T-by-type.p2.zoomclick-d3.icon.n-apps\endcsname{10}
\expandafter\def\csname pct@rev.desc.d7.sspro.tau-T-by-type.p2.zoomclick-d3.icon.n-apps\endcsname{1000.0}
\expandafter\def\csname ppt@rev.desc.d7.sspro.tau-T-by-type.p2.zoomclick-d3.icon.n-apps\endcsname{$+$1000.0}
\expandafter\def\csname val@rev.desc.d7.sspro.tau-T-by-type.p2.zoomclick-d3.icon.n-tasks\endcsname{28}
\expandafter\def\csname cnt@rev.desc.d7.sspro.tau-T-by-type.p2.zoomclick-d3.icon.n-tasks\endcsname{28}
\expandafter\def\csname pct@rev.desc.d7.sspro.tau-T-by-type.p2.zoomclick-d3.icon.n-tasks\endcsname{2800.0}
\expandafter\def\csname ppt@rev.desc.d7.sspro.tau-T-by-type.p2.zoomclick-d3.icon.n-tasks\endcsname{$+$2800.0}
\expandafter\def\csname val@rev.desc.d7.sspro.tau-T-by-type.p2.zoomclick-d3.icon.tau.ci90.0\endcsname{0.127}
\expandafter\def\csname cnt@rev.desc.d7.sspro.tau-T-by-type.p2.zoomclick-d3.icon.tau.ci90.0\endcsname{0}
\expandafter\def\csname pct@rev.desc.d7.sspro.tau-T-by-type.p2.zoomclick-d3.icon.tau.ci90.0\endcsname{12.7}
\expandafter\def\csname ppt@rev.desc.d7.sspro.tau-T-by-type.p2.zoomclick-d3.icon.tau.ci90.0\endcsname{$+$12.7}
\expandafter\def\csname val@rev.desc.d7.sspro.tau-T-by-type.p2.zoomclick-d3.icon.tau.ci90.1\endcsname{0.460}
\expandafter\def\csname cnt@rev.desc.d7.sspro.tau-T-by-type.p2.zoomclick-d3.icon.tau.ci90.1\endcsname{0}
\expandafter\def\csname pct@rev.desc.d7.sspro.tau-T-by-type.p2.zoomclick-d3.icon.tau.ci90.1\endcsname{46.0}
\expandafter\def\csname ppt@rev.desc.d7.sspro.tau-T-by-type.p2.zoomclick-d3.icon.tau.ci90.1\endcsname{$+$46.0}
\expandafter\def\csname val@rev.desc.d7.sspro.tau-T-by-type.p2.zoomclick-d3.icon.tau.ci95-excludes-0\endcsname{1}
\expandafter\def\csname cnt@rev.desc.d7.sspro.tau-T-by-type.p2.zoomclick-d3.icon.tau.ci95-excludes-0\endcsname{1}
\expandafter\def\csname pct@rev.desc.d7.sspro.tau-T-by-type.p2.zoomclick-d3.icon.tau.ci95-excludes-0\endcsname{100.0}
\expandafter\def\csname ppt@rev.desc.d7.sspro.tau-T-by-type.p2.zoomclick-d3.icon.tau.ci95-excludes-0\endcsname{$+$100.0}
\expandafter\def\csname val@rev.desc.d7.sspro.tau-T-by-type.p2.zoomclick-d3.icon.tau.ci95.0\endcsname{0.107}
\expandafter\def\csname cnt@rev.desc.d7.sspro.tau-T-by-type.p2.zoomclick-d3.icon.tau.ci95.0\endcsname{0}
\expandafter\def\csname pct@rev.desc.d7.sspro.tau-T-by-type.p2.zoomclick-d3.icon.tau.ci95.0\endcsname{10.7}
\expandafter\def\csname ppt@rev.desc.d7.sspro.tau-T-by-type.p2.zoomclick-d3.icon.tau.ci95.0\endcsname{$+$10.7}
\expandafter\def\csname val@rev.desc.d7.sspro.tau-T-by-type.p2.zoomclick-d3.icon.tau.ci95.1\endcsname{0.500}
\expandafter\def\csname cnt@rev.desc.d7.sspro.tau-T-by-type.p2.zoomclick-d3.icon.tau.ci95.1\endcsname{0}
\expandafter\def\csname pct@rev.desc.d7.sspro.tau-T-by-type.p2.zoomclick-d3.icon.tau.ci95.1\endcsname{50.0}
\expandafter\def\csname ppt@rev.desc.d7.sspro.tau-T-by-type.p2.zoomclick-d3.icon.tau.ci95.1\endcsname{$+$50.0}
\expandafter\def\csname val@rev.desc.d7.sspro.tau-T-by-type.p2.zoomclick-d3.icon.tau.mean\endcsname{0.290}
\expandafter\def\csname cnt@rev.desc.d7.sspro.tau-T-by-type.p2.zoomclick-d3.icon.tau.mean\endcsname{0}
\expandafter\def\csname pct@rev.desc.d7.sspro.tau-T-by-type.p2.zoomclick-d3.icon.tau.mean\endcsname{29.0}
\expandafter\def\csname ppt@rev.desc.d7.sspro.tau-T-by-type.p2.zoomclick-d3.icon.tau.mean\endcsname{$+$29.0}
\expandafter\def\csname val@rev.desc.d7.sspro.tau-T-by-type.p2.zoomclick-d3.icon.tau.n-apps\endcsname{10}
\expandafter\def\csname cnt@rev.desc.d7.sspro.tau-T-by-type.p2.zoomclick-d3.icon.tau.n-apps\endcsname{10}
\expandafter\def\csname pct@rev.desc.d7.sspro.tau-T-by-type.p2.zoomclick-d3.icon.tau.n-apps\endcsname{1000.0}
\expandafter\def\csname ppt@rev.desc.d7.sspro.tau-T-by-type.p2.zoomclick-d3.icon.tau.n-apps\endcsname{$+$1000.0}
\expandafter\def\csname val@rev.desc.d7.sspro.tau-T-by-type.p2.zoomclick-d3.icon.tau.tost1.ci90-inside\endcsname{0}
\expandafter\def\csname cnt@rev.desc.d7.sspro.tau-T-by-type.p2.zoomclick-d3.icon.tau.tost1.ci90-inside\endcsname{0}
\expandafter\def\csname pct@rev.desc.d7.sspro.tau-T-by-type.p2.zoomclick-d3.icon.tau.tost1.ci90-inside\endcsname{0.0}
\expandafter\def\csname ppt@rev.desc.d7.sspro.tau-T-by-type.p2.zoomclick-d3.icon.tau.tost1.ci90-inside\endcsname{0.0}
\expandafter\def\csname val@rev.desc.d7.sspro.tau-T-by-type.p2.zoomclick-d3.icon.tau.tost1.equivalent\endcsname{0}
\expandafter\def\csname cnt@rev.desc.d7.sspro.tau-T-by-type.p2.zoomclick-d3.icon.tau.tost1.equivalent\endcsname{0}
\expandafter\def\csname pct@rev.desc.d7.sspro.tau-T-by-type.p2.zoomclick-d3.icon.tau.tost1.equivalent\endcsname{0.0}
\expandafter\def\csname ppt@rev.desc.d7.sspro.tau-T-by-type.p2.zoomclick-d3.icon.tau.tost1.equivalent\endcsname{0.0}
\expandafter\def\csname val@rev.desc.d7.sspro.tau-T-by-type.p2.zoomclick-d3.icon.tau.tost1.p\endcsname{1.000}
\expandafter\def\csname cnt@rev.desc.d7.sspro.tau-T-by-type.p2.zoomclick-d3.icon.tau.tost1.p\endcsname{1}
\expandafter\def\csname pct@rev.desc.d7.sspro.tau-T-by-type.p2.zoomclick-d3.icon.tau.tost1.p\endcsname{100.0}
\expandafter\def\csname ppt@rev.desc.d7.sspro.tau-T-by-type.p2.zoomclick-d3.icon.tau.tost1.p\endcsname{$+$100.0}
\expandafter\def\csname val@rev.desc.d7.sspro.tau-T-by-type.p2.zoomclick-d3.icon.tau.tost5.ci90-inside\endcsname{0}
\expandafter\def\csname cnt@rev.desc.d7.sspro.tau-T-by-type.p2.zoomclick-d3.icon.tau.tost5.ci90-inside\endcsname{0}
\expandafter\def\csname pct@rev.desc.d7.sspro.tau-T-by-type.p2.zoomclick-d3.icon.tau.tost5.ci90-inside\endcsname{0.0}
\expandafter\def\csname ppt@rev.desc.d7.sspro.tau-T-by-type.p2.zoomclick-d3.icon.tau.tost5.ci90-inside\endcsname{0.0}
\expandafter\def\csname val@rev.desc.d7.sspro.tau-T-by-type.p2.zoomclick-d3.icon.tau.tost5.equivalent\endcsname{0}
\expandafter\def\csname cnt@rev.desc.d7.sspro.tau-T-by-type.p2.zoomclick-d3.icon.tau.tost5.equivalent\endcsname{0}
\expandafter\def\csname pct@rev.desc.d7.sspro.tau-T-by-type.p2.zoomclick-d3.icon.tau.tost5.equivalent\endcsname{0.0}
\expandafter\def\csname ppt@rev.desc.d7.sspro.tau-T-by-type.p2.zoomclick-d3.icon.tau.tost5.equivalent\endcsname{0.0}
\expandafter\def\csname val@rev.desc.d7.sspro.tau-T-by-type.p2.zoomclick-d3.icon.tau.tost5.p\endcsname{0.998}
\expandafter\def\csname cnt@rev.desc.d7.sspro.tau-T-by-type.p2.zoomclick-d3.icon.tau.tost5.p\endcsname{1}
\expandafter\def\csname pct@rev.desc.d7.sspro.tau-T-by-type.p2.zoomclick-d3.icon.tau.tost5.p\endcsname{99.8}
\expandafter\def\csname ppt@rev.desc.d7.sspro.tau-T-by-type.p2.zoomclick-d3.icon.tau.tost5.p\endcsname{$+$99.8}
\expandafter\def\csname val@rev.desc.d7.sspro.tau-T-by-type.p2.zoomclick-d3.text.acc-F\endcsname{0.971}
\expandafter\def\csname cnt@rev.desc.d7.sspro.tau-T-by-type.p2.zoomclick-d3.text.acc-F\endcsname{1}
\expandafter\def\csname pct@rev.desc.d7.sspro.tau-T-by-type.p2.zoomclick-d3.text.acc-F\endcsname{97.1}
\expandafter\def\csname ppt@rev.desc.d7.sspro.tau-T-by-type.p2.zoomclick-d3.text.acc-F\endcsname{$+$97.1}
\expandafter\def\csname val@rev.desc.d7.sspro.tau-T-by-type.p2.zoomclick-d3.text.acc-direct\endcsname{0.838}
\expandafter\def\csname cnt@rev.desc.d7.sspro.tau-T-by-type.p2.zoomclick-d3.text.acc-direct\endcsname{1}
\expandafter\def\csname pct@rev.desc.d7.sspro.tau-T-by-type.p2.zoomclick-d3.text.acc-direct\endcsname{83.8}
\expandafter\def\csname ppt@rev.desc.d7.sspro.tau-T-by-type.p2.zoomclick-d3.text.acc-direct\endcsname{$+$83.8}
\expandafter\def\csname val@rev.desc.d7.sspro.tau-T-by-type.p2.zoomclick-d3.text.n-apps\endcsname{12}
\expandafter\def\csname cnt@rev.desc.d7.sspro.tau-T-by-type.p2.zoomclick-d3.text.n-apps\endcsname{12}
\expandafter\def\csname pct@rev.desc.d7.sspro.tau-T-by-type.p2.zoomclick-d3.text.n-apps\endcsname{1200.0}
\expandafter\def\csname ppt@rev.desc.d7.sspro.tau-T-by-type.p2.zoomclick-d3.text.n-apps\endcsname{$+$1200.0}
\expandafter\def\csname val@rev.desc.d7.sspro.tau-T-by-type.p2.zoomclick-d3.text.n-tasks\endcsname{68}
\expandafter\def\csname cnt@rev.desc.d7.sspro.tau-T-by-type.p2.zoomclick-d3.text.n-tasks\endcsname{68}
\expandafter\def\csname pct@rev.desc.d7.sspro.tau-T-by-type.p2.zoomclick-d3.text.n-tasks\endcsname{6800.0}
\expandafter\def\csname ppt@rev.desc.d7.sspro.tau-T-by-type.p2.zoomclick-d3.text.n-tasks\endcsname{$+$6800.0}
\expandafter\def\csname val@rev.desc.d7.sspro.tau-T-by-type.p2.zoomclick-d3.text.tau.ci90.0\endcsname{0.031}
\expandafter\def\csname cnt@rev.desc.d7.sspro.tau-T-by-type.p2.zoomclick-d3.text.tau.ci90.0\endcsname{0}
\expandafter\def\csname pct@rev.desc.d7.sspro.tau-T-by-type.p2.zoomclick-d3.text.tau.ci90.0\endcsname{3.1}
\expandafter\def\csname ppt@rev.desc.d7.sspro.tau-T-by-type.p2.zoomclick-d3.text.tau.ci90.0\endcsname{$+$3.1}
\expandafter\def\csname val@rev.desc.d7.sspro.tau-T-by-type.p2.zoomclick-d3.text.tau.ci90.1\endcsname{0.173}
\expandafter\def\csname cnt@rev.desc.d7.sspro.tau-T-by-type.p2.zoomclick-d3.text.tau.ci90.1\endcsname{0}
\expandafter\def\csname pct@rev.desc.d7.sspro.tau-T-by-type.p2.zoomclick-d3.text.tau.ci90.1\endcsname{17.3}
\expandafter\def\csname ppt@rev.desc.d7.sspro.tau-T-by-type.p2.zoomclick-d3.text.tau.ci90.1\endcsname{$+$17.3}
\expandafter\def\csname val@rev.desc.d7.sspro.tau-T-by-type.p2.zoomclick-d3.text.tau.ci95-excludes-0\endcsname{1}
\expandafter\def\csname cnt@rev.desc.d7.sspro.tau-T-by-type.p2.zoomclick-d3.text.tau.ci95-excludes-0\endcsname{1}
\expandafter\def\csname pct@rev.desc.d7.sspro.tau-T-by-type.p2.zoomclick-d3.text.tau.ci95-excludes-0\endcsname{100.0}
\expandafter\def\csname ppt@rev.desc.d7.sspro.tau-T-by-type.p2.zoomclick-d3.text.tau.ci95-excludes-0\endcsname{$+$100.0}
\expandafter\def\csname val@rev.desc.d7.sspro.tau-T-by-type.p2.zoomclick-d3.text.tau.ci95.0\endcsname{0.024}
\expandafter\def\csname cnt@rev.desc.d7.sspro.tau-T-by-type.p2.zoomclick-d3.text.tau.ci95.0\endcsname{0}
\expandafter\def\csname pct@rev.desc.d7.sspro.tau-T-by-type.p2.zoomclick-d3.text.tau.ci95.0\endcsname{2.4}
\expandafter\def\csname ppt@rev.desc.d7.sspro.tau-T-by-type.p2.zoomclick-d3.text.tau.ci95.0\endcsname{$+$2.4}
\expandafter\def\csname val@rev.desc.d7.sspro.tau-T-by-type.p2.zoomclick-d3.text.tau.ci95.1\endcsname{0.190}
\expandafter\def\csname cnt@rev.desc.d7.sspro.tau-T-by-type.p2.zoomclick-d3.text.tau.ci95.1\endcsname{0}
\expandafter\def\csname pct@rev.desc.d7.sspro.tau-T-by-type.p2.zoomclick-d3.text.tau.ci95.1\endcsname{19.0}
\expandafter\def\csname ppt@rev.desc.d7.sspro.tau-T-by-type.p2.zoomclick-d3.text.tau.ci95.1\endcsname{$+$19.0}
\expandafter\def\csname val@rev.desc.d7.sspro.tau-T-by-type.p2.zoomclick-d3.text.tau.mean\endcsname{0.100}
\expandafter\def\csname cnt@rev.desc.d7.sspro.tau-T-by-type.p2.zoomclick-d3.text.tau.mean\endcsname{0}
\expandafter\def\csname pct@rev.desc.d7.sspro.tau-T-by-type.p2.zoomclick-d3.text.tau.mean\endcsname{10.0}
\expandafter\def\csname ppt@rev.desc.d7.sspro.tau-T-by-type.p2.zoomclick-d3.text.tau.mean\endcsname{$+$10.0}
\expandafter\def\csname val@rev.desc.d7.sspro.tau-T-by-type.p2.zoomclick-d3.text.tau.n-apps\endcsname{12}
\expandafter\def\csname cnt@rev.desc.d7.sspro.tau-T-by-type.p2.zoomclick-d3.text.tau.n-apps\endcsname{12}
\expandafter\def\csname pct@rev.desc.d7.sspro.tau-T-by-type.p2.zoomclick-d3.text.tau.n-apps\endcsname{1200.0}
\expandafter\def\csname ppt@rev.desc.d7.sspro.tau-T-by-type.p2.zoomclick-d3.text.tau.n-apps\endcsname{$+$1200.0}
\expandafter\def\csname val@rev.desc.d7.sspro.tau-T-by-type.p2.zoomclick-d3.text.tau.tost1.ci90-inside\endcsname{0}
\expandafter\def\csname cnt@rev.desc.d7.sspro.tau-T-by-type.p2.zoomclick-d3.text.tau.tost1.ci90-inside\endcsname{0}
\expandafter\def\csname pct@rev.desc.d7.sspro.tau-T-by-type.p2.zoomclick-d3.text.tau.tost1.ci90-inside\endcsname{0.0}
\expandafter\def\csname ppt@rev.desc.d7.sspro.tau-T-by-type.p2.zoomclick-d3.text.tau.tost1.ci90-inside\endcsname{0.0}
\expandafter\def\csname val@rev.desc.d7.sspro.tau-T-by-type.p2.zoomclick-d3.text.tau.tost1.equivalent\endcsname{0}
\expandafter\def\csname cnt@rev.desc.d7.sspro.tau-T-by-type.p2.zoomclick-d3.text.tau.tost1.equivalent\endcsname{0}
\expandafter\def\csname pct@rev.desc.d7.sspro.tau-T-by-type.p2.zoomclick-d3.text.tau.tost1.equivalent\endcsname{0.0}
\expandafter\def\csname ppt@rev.desc.d7.sspro.tau-T-by-type.p2.zoomclick-d3.text.tau.tost1.equivalent\endcsname{0.0}
\expandafter\def\csname val@rev.desc.d7.sspro.tau-T-by-type.p2.zoomclick-d3.text.tau.tost1.p\endcsname{0.991}
\expandafter\def\csname cnt@rev.desc.d7.sspro.tau-T-by-type.p2.zoomclick-d3.text.tau.tost1.p\endcsname{1}
\expandafter\def\csname pct@rev.desc.d7.sspro.tau-T-by-type.p2.zoomclick-d3.text.tau.tost1.p\endcsname{99.1}
\expandafter\def\csname ppt@rev.desc.d7.sspro.tau-T-by-type.p2.zoomclick-d3.text.tau.tost1.p\endcsname{$+$99.1}
\expandafter\def\csname val@rev.desc.d7.sspro.tau-T-by-type.p2.zoomclick-d3.text.tau.tost5.ci90-inside\endcsname{0}
\expandafter\def\csname cnt@rev.desc.d7.sspro.tau-T-by-type.p2.zoomclick-d3.text.tau.tost5.ci90-inside\endcsname{0}
\expandafter\def\csname pct@rev.desc.d7.sspro.tau-T-by-type.p2.zoomclick-d3.text.tau.tost5.ci90-inside\endcsname{0.0}
\expandafter\def\csname ppt@rev.desc.d7.sspro.tau-T-by-type.p2.zoomclick-d3.text.tau.tost5.ci90-inside\endcsname{0.0}
\expandafter\def\csname val@rev.desc.d7.sspro.tau-T-by-type.p2.zoomclick-d3.text.tau.tost5.equivalent\endcsname{0}
\expandafter\def\csname cnt@rev.desc.d7.sspro.tau-T-by-type.p2.zoomclick-d3.text.tau.tost5.equivalent\endcsname{0}
\expandafter\def\csname pct@rev.desc.d7.sspro.tau-T-by-type.p2.zoomclick-d3.text.tau.tost5.equivalent\endcsname{0.0}
\expandafter\def\csname ppt@rev.desc.d7.sspro.tau-T-by-type.p2.zoomclick-d3.text.tau.tost5.equivalent\endcsname{0.0}
\expandafter\def\csname val@rev.desc.d7.sspro.tau-T-by-type.p2.zoomclick-d3.text.tau.tost5.p\endcsname{0.885}
\expandafter\def\csname cnt@rev.desc.d7.sspro.tau-T-by-type.p2.zoomclick-d3.text.tau.tost5.p\endcsname{1}
\expandafter\def\csname pct@rev.desc.d7.sspro.tau-T-by-type.p2.zoomclick-d3.text.tau.tost5.p\endcsname{88.5}
\expandafter\def\csname ppt@rev.desc.d7.sspro.tau-T-by-type.p2.zoomclick-d3.text.tau.tost5.p\endcsname{$+$88.5}
\expandafter\def\csname val@rev.desc.d7.uiv.composition.A.n\endcsname{64}
\expandafter\def\csname cnt@rev.desc.d7.uiv.composition.A.n\endcsname{64}
\expandafter\def\csname pct@rev.desc.d7.uiv.composition.A.n\endcsname{6400.0}
\expandafter\def\csname ppt@rev.desc.d7.uiv.composition.A.n\endcsname{$+$6400.0}
\expandafter\def\csname val@rev.desc.d7.uiv.composition.A.n-icon\endcsname{41}
\expandafter\def\csname cnt@rev.desc.d7.uiv.composition.A.n-icon\endcsname{41}
\expandafter\def\csname pct@rev.desc.d7.uiv.composition.A.n-icon\endcsname{4100.0}
\expandafter\def\csname ppt@rev.desc.d7.uiv.composition.A.n-icon\endcsname{$+$4100.0}
\expandafter\def\csname val@rev.desc.d7.uiv.composition.A.n-text\endcsname{23}
\expandafter\def\csname cnt@rev.desc.d7.uiv.composition.A.n-text\endcsname{23}
\expandafter\def\csname pct@rev.desc.d7.uiv.composition.A.n-text\endcsname{2300.0}
\expandafter\def\csname ppt@rev.desc.d7.uiv.composition.A.n-text\endcsname{$+$2300.0}
\expandafter\def\csname val@rev.desc.d7.uiv.composition.A.share-icon\endcsname{0.641}
\expandafter\def\csname cnt@rev.desc.d7.uiv.composition.A.share-icon\endcsname{1}
\expandafter\def\csname pct@rev.desc.d7.uiv.composition.A.share-icon\endcsname{64.1}
\expandafter\def\csname ppt@rev.desc.d7.uiv.composition.A.share-icon\endcsname{$+$64.1}
\expandafter\def\csname val@rev.desc.d7.uiv.composition.A.share-text\endcsname{0.359}
\expandafter\def\csname cnt@rev.desc.d7.uiv.composition.A.share-text\endcsname{0}
\expandafter\def\csname pct@rev.desc.d7.uiv.composition.A.share-text\endcsname{35.9}
\expandafter\def\csname ppt@rev.desc.d7.uiv.composition.A.share-text\endcsname{$+$35.9}
\expandafter\def\csname val@rev.desc.d7.uiv.composition.C.n\endcsname{96}
\expandafter\def\csname cnt@rev.desc.d7.uiv.composition.C.n\endcsname{96}
\expandafter\def\csname pct@rev.desc.d7.uiv.composition.C.n\endcsname{9600.0}
\expandafter\def\csname ppt@rev.desc.d7.uiv.composition.C.n\endcsname{$+$9600.0}
\expandafter\def\csname val@rev.desc.d7.uiv.composition.C.n-icon\endcsname{80}
\expandafter\def\csname cnt@rev.desc.d7.uiv.composition.C.n-icon\endcsname{80}
\expandafter\def\csname pct@rev.desc.d7.uiv.composition.C.n-icon\endcsname{8000.0}
\expandafter\def\csname ppt@rev.desc.d7.uiv.composition.C.n-icon\endcsname{$+$8000.0}
\expandafter\def\csname val@rev.desc.d7.uiv.composition.C.n-text\endcsname{16}
\expandafter\def\csname cnt@rev.desc.d7.uiv.composition.C.n-text\endcsname{16}
\expandafter\def\csname pct@rev.desc.d7.uiv.composition.C.n-text\endcsname{1600.0}
\expandafter\def\csname ppt@rev.desc.d7.uiv.composition.C.n-text\endcsname{$+$1600.0}
\expandafter\def\csname val@rev.desc.d7.uiv.composition.C.share-icon\endcsname{0.833}
\expandafter\def\csname cnt@rev.desc.d7.uiv.composition.C.share-icon\endcsname{1}
\expandafter\def\csname pct@rev.desc.d7.uiv.composition.C.share-icon\endcsname{83.3}
\expandafter\def\csname ppt@rev.desc.d7.uiv.composition.C.share-icon\endcsname{$+$83.3}
\expandafter\def\csname val@rev.desc.d7.uiv.composition.C.share-text\endcsname{0.167}
\expandafter\def\csname cnt@rev.desc.d7.uiv.composition.C.share-text\endcsname{0}
\expandafter\def\csname pct@rev.desc.d7.uiv.composition.C.share-text\endcsname{16.7}
\expandafter\def\csname ppt@rev.desc.d7.uiv.composition.C.share-text\endcsname{$+$16.7}
\expandafter\def\csname val@rev.desc.d7.uiv.composition.T.n\endcsname{144}
\expandafter\def\csname cnt@rev.desc.d7.uiv.composition.T.n\endcsname{144}
\expandafter\def\csname pct@rev.desc.d7.uiv.composition.T.n\endcsname{14400.0}
\expandafter\def\csname ppt@rev.desc.d7.uiv.composition.T.n\endcsname{$+$14400.0}
\expandafter\def\csname val@rev.desc.d7.uiv.composition.T.n-icon\endcsname{114}
\expandafter\def\csname cnt@rev.desc.d7.uiv.composition.T.n-icon\endcsname{114}
\expandafter\def\csname pct@rev.desc.d7.uiv.composition.T.n-icon\endcsname{11400.0}
\expandafter\def\csname ppt@rev.desc.d7.uiv.composition.T.n-icon\endcsname{$+$11400.0}
\expandafter\def\csname val@rev.desc.d7.uiv.composition.T.n-text\endcsname{30}
\expandafter\def\csname cnt@rev.desc.d7.uiv.composition.T.n-text\endcsname{30}
\expandafter\def\csname pct@rev.desc.d7.uiv.composition.T.n-text\endcsname{3000.0}
\expandafter\def\csname ppt@rev.desc.d7.uiv.composition.T.n-text\endcsname{$+$3000.0}
\expandafter\def\csname val@rev.desc.d7.uiv.composition.T.share-icon\endcsname{0.792}
\expandafter\def\csname cnt@rev.desc.d7.uiv.composition.T.share-icon\endcsname{1}
\expandafter\def\csname pct@rev.desc.d7.uiv.composition.T.share-icon\endcsname{79.2}
\expandafter\def\csname ppt@rev.desc.d7.uiv.composition.T.share-icon\endcsname{$+$79.2}
\expandafter\def\csname val@rev.desc.d7.uiv.composition.T.share-text\endcsname{0.208}
\expandafter\def\csname cnt@rev.desc.d7.uiv.composition.T.share-text\endcsname{0}
\expandafter\def\csname pct@rev.desc.d7.uiv.composition.T.share-text\endcsname{20.8}
\expandafter\def\csname ppt@rev.desc.d7.uiv.composition.T.share-text\endcsname{$+$20.8}
\expandafter\def\csname val@rev.desc.d7.uiv.tau-T-by-type.p1.center-d1.icon-minus-text-on-apps-with-both.ci90.0\endcsname{0.000}
\expandafter\def\csname cnt@rev.desc.d7.uiv.tau-T-by-type.p1.center-d1.icon-minus-text-on-apps-with-both.ci90.0\endcsname{0}
\expandafter\def\csname pct@rev.desc.d7.uiv.tau-T-by-type.p1.center-d1.icon-minus-text-on-apps-with-both.ci90.0\endcsname{0.0}
\expandafter\def\csname ppt@rev.desc.d7.uiv.tau-T-by-type.p1.center-d1.icon-minus-text-on-apps-with-both.ci90.0\endcsname{0.0}
\expandafter\def\csname val@rev.desc.d7.uiv.tau-T-by-type.p1.center-d1.icon-minus-text-on-apps-with-both.ci90.1\endcsname{0.204}
\expandafter\def\csname cnt@rev.desc.d7.uiv.tau-T-by-type.p1.center-d1.icon-minus-text-on-apps-with-both.ci90.1\endcsname{0}
\expandafter\def\csname pct@rev.desc.d7.uiv.tau-T-by-type.p1.center-d1.icon-minus-text-on-apps-with-both.ci90.1\endcsname{20.4}
\expandafter\def\csname ppt@rev.desc.d7.uiv.tau-T-by-type.p1.center-d1.icon-minus-text-on-apps-with-both.ci90.1\endcsname{$+$20.4}
\expandafter\def\csname val@rev.desc.d7.uiv.tau-T-by-type.p1.center-d1.icon-minus-text-on-apps-with-both.ci95-excludes-0\endcsname{0}
\expandafter\def\csname cnt@rev.desc.d7.uiv.tau-T-by-type.p1.center-d1.icon-minus-text-on-apps-with-both.ci95-excludes-0\endcsname{0}
\expandafter\def\csname pct@rev.desc.d7.uiv.tau-T-by-type.p1.center-d1.icon-minus-text-on-apps-with-both.ci95-excludes-0\endcsname{0.0}
\expandafter\def\csname ppt@rev.desc.d7.uiv.tau-T-by-type.p1.center-d1.icon-minus-text-on-apps-with-both.ci95-excludes-0\endcsname{0.0}
\expandafter\def\csname val@rev.desc.d7.uiv.tau-T-by-type.p1.center-d1.icon-minus-text-on-apps-with-both.ci95.0\endcsname{-0.011}
\expandafter\def\csname cnt@rev.desc.d7.uiv.tau-T-by-type.p1.center-d1.icon-minus-text-on-apps-with-both.ci95.0\endcsname{0}
\expandafter\def\csname pct@rev.desc.d7.uiv.tau-T-by-type.p1.center-d1.icon-minus-text-on-apps-with-both.ci95.0\endcsname{$-$1.1}
\expandafter\def\csname ppt@rev.desc.d7.uiv.tau-T-by-type.p1.center-d1.icon-minus-text-on-apps-with-both.ci95.0\endcsname{$-$1.1}
\expandafter\def\csname val@rev.desc.d7.uiv.tau-T-by-type.p1.center-d1.icon-minus-text-on-apps-with-both.ci95.1\endcsname{0.229}
\expandafter\def\csname cnt@rev.desc.d7.uiv.tau-T-by-type.p1.center-d1.icon-minus-text-on-apps-with-both.ci95.1\endcsname{0}
\expandafter\def\csname pct@rev.desc.d7.uiv.tau-T-by-type.p1.center-d1.icon-minus-text-on-apps-with-both.ci95.1\endcsname{22.9}
\expandafter\def\csname ppt@rev.desc.d7.uiv.tau-T-by-type.p1.center-d1.icon-minus-text-on-apps-with-both.ci95.1\endcsname{$+$22.9}
\expandafter\def\csname val@rev.desc.d7.uiv.tau-T-by-type.p1.center-d1.icon-minus-text-on-apps-with-both.mean\endcsname{0.089}
\expandafter\def\csname cnt@rev.desc.d7.uiv.tau-T-by-type.p1.center-d1.icon-minus-text-on-apps-with-both.mean\endcsname{0}
\expandafter\def\csname pct@rev.desc.d7.uiv.tau-T-by-type.p1.center-d1.icon-minus-text-on-apps-with-both.mean\endcsname{8.9}
\expandafter\def\csname ppt@rev.desc.d7.uiv.tau-T-by-type.p1.center-d1.icon-minus-text-on-apps-with-both.mean\endcsname{$+$8.9}
\expandafter\def\csname val@rev.desc.d7.uiv.tau-T-by-type.p1.center-d1.icon-minus-text-on-apps-with-both.n-apps\endcsname{14}
\expandafter\def\csname cnt@rev.desc.d7.uiv.tau-T-by-type.p1.center-d1.icon-minus-text-on-apps-with-both.n-apps\endcsname{14}
\expandafter\def\csname pct@rev.desc.d7.uiv.tau-T-by-type.p1.center-d1.icon-minus-text-on-apps-with-both.n-apps\endcsname{1400.0}
\expandafter\def\csname ppt@rev.desc.d7.uiv.tau-T-by-type.p1.center-d1.icon-minus-text-on-apps-with-both.n-apps\endcsname{$+$1400.0}
\expandafter\def\csname val@rev.desc.d7.uiv.tau-T-by-type.p1.center-d1.icon-minus-text-on-apps-with-both.tost1.ci90-inside\endcsname{0}
\expandafter\def\csname cnt@rev.desc.d7.uiv.tau-T-by-type.p1.center-d1.icon-minus-text-on-apps-with-both.tost1.ci90-inside\endcsname{0}
\expandafter\def\csname pct@rev.desc.d7.uiv.tau-T-by-type.p1.center-d1.icon-minus-text-on-apps-with-both.tost1.ci90-inside\endcsname{0.0}
\expandafter\def\csname ppt@rev.desc.d7.uiv.tau-T-by-type.p1.center-d1.icon-minus-text-on-apps-with-both.tost1.ci90-inside\endcsname{0.0}
\expandafter\def\csname val@rev.desc.d7.uiv.tau-T-by-type.p1.center-d1.icon-minus-text-on-apps-with-both.tost1.equivalent\endcsname{0}
\expandafter\def\csname cnt@rev.desc.d7.uiv.tau-T-by-type.p1.center-d1.icon-minus-text-on-apps-with-both.tost1.equivalent\endcsname{0}
\expandafter\def\csname pct@rev.desc.d7.uiv.tau-T-by-type.p1.center-d1.icon-minus-text-on-apps-with-both.tost1.equivalent\endcsname{0.0}
\expandafter\def\csname ppt@rev.desc.d7.uiv.tau-T-by-type.p1.center-d1.icon-minus-text-on-apps-with-both.tost1.equivalent\endcsname{0.0}
\expandafter\def\csname val@rev.desc.d7.uiv.tau-T-by-type.p1.center-d1.icon-minus-text-on-apps-with-both.tost1.p\endcsname{0.924}
\expandafter\def\csname cnt@rev.desc.d7.uiv.tau-T-by-type.p1.center-d1.icon-minus-text-on-apps-with-both.tost1.p\endcsname{1}
\expandafter\def\csname pct@rev.desc.d7.uiv.tau-T-by-type.p1.center-d1.icon-minus-text-on-apps-with-both.tost1.p\endcsname{92.4}
\expandafter\def\csname ppt@rev.desc.d7.uiv.tau-T-by-type.p1.center-d1.icon-minus-text-on-apps-with-both.tost1.p\endcsname{$+$92.4}
\expandafter\def\csname val@rev.desc.d7.uiv.tau-T-by-type.p1.center-d1.icon-minus-text-on-apps-with-both.tost5.ci90-inside\endcsname{0}
\expandafter\def\csname cnt@rev.desc.d7.uiv.tau-T-by-type.p1.center-d1.icon-minus-text-on-apps-with-both.tost5.ci90-inside\endcsname{0}
\expandafter\def\csname pct@rev.desc.d7.uiv.tau-T-by-type.p1.center-d1.icon-minus-text-on-apps-with-both.tost5.ci90-inside\endcsname{0.0}
\expandafter\def\csname ppt@rev.desc.d7.uiv.tau-T-by-type.p1.center-d1.icon-minus-text-on-apps-with-both.tost5.ci90-inside\endcsname{0.0}
\expandafter\def\csname val@rev.desc.d7.uiv.tau-T-by-type.p1.center-d1.icon-minus-text-on-apps-with-both.tost5.equivalent\endcsname{0}
\expandafter\def\csname cnt@rev.desc.d7.uiv.tau-T-by-type.p1.center-d1.icon-minus-text-on-apps-with-both.tost5.equivalent\endcsname{0}
\expandafter\def\csname pct@rev.desc.d7.uiv.tau-T-by-type.p1.center-d1.icon-minus-text-on-apps-with-both.tost5.equivalent\endcsname{0.0}
\expandafter\def\csname ppt@rev.desc.d7.uiv.tau-T-by-type.p1.center-d1.icon-minus-text-on-apps-with-both.tost5.equivalent\endcsname{0.0}
\expandafter\def\csname val@rev.desc.d7.uiv.tau-T-by-type.p1.center-d1.icon-minus-text-on-apps-with-both.tost5.p\endcsname{0.701}
\expandafter\def\csname cnt@rev.desc.d7.uiv.tau-T-by-type.p1.center-d1.icon-minus-text-on-apps-with-both.tost5.p\endcsname{1}
\expandafter\def\csname pct@rev.desc.d7.uiv.tau-T-by-type.p1.center-d1.icon-minus-text-on-apps-with-both.tost5.p\endcsname{70.1}
\expandafter\def\csname ppt@rev.desc.d7.uiv.tau-T-by-type.p1.center-d1.icon-minus-text-on-apps-with-both.tost5.p\endcsname{$+$70.1}
\expandafter\def\csname val@rev.desc.d7.uiv.tau-T-by-type.p1.center-d1.icon.acc-F\endcsname{0.781}
\expandafter\def\csname cnt@rev.desc.d7.uiv.tau-T-by-type.p1.center-d1.icon.acc-F\endcsname{1}
\expandafter\def\csname pct@rev.desc.d7.uiv.tau-T-by-type.p1.center-d1.icon.acc-F\endcsname{78.1}
\expandafter\def\csname ppt@rev.desc.d7.uiv.tau-T-by-type.p1.center-d1.icon.acc-F\endcsname{$+$78.1}
\expandafter\def\csname val@rev.desc.d7.uiv.tau-T-by-type.p1.center-d1.icon.acc-direct\endcsname{0.825}
\expandafter\def\csname cnt@rev.desc.d7.uiv.tau-T-by-type.p1.center-d1.icon.acc-direct\endcsname{1}
\expandafter\def\csname pct@rev.desc.d7.uiv.tau-T-by-type.p1.center-d1.icon.acc-direct\endcsname{82.5}
\expandafter\def\csname ppt@rev.desc.d7.uiv.tau-T-by-type.p1.center-d1.icon.acc-direct\endcsname{$+$82.5}
\expandafter\def\csname val@rev.desc.d7.uiv.tau-T-by-type.p1.center-d1.icon.n-apps\endcsname{23}
\expandafter\def\csname cnt@rev.desc.d7.uiv.tau-T-by-type.p1.center-d1.icon.n-apps\endcsname{23}
\expandafter\def\csname pct@rev.desc.d7.uiv.tau-T-by-type.p1.center-d1.icon.n-apps\endcsname{2300.0}
\expandafter\def\csname ppt@rev.desc.d7.uiv.tau-T-by-type.p1.center-d1.icon.n-apps\endcsname{$+$2300.0}
\expandafter\def\csname val@rev.desc.d7.uiv.tau-T-by-type.p1.center-d1.icon.n-tasks\endcsname{114}
\expandafter\def\csname cnt@rev.desc.d7.uiv.tau-T-by-type.p1.center-d1.icon.n-tasks\endcsname{114}
\expandafter\def\csname pct@rev.desc.d7.uiv.tau-T-by-type.p1.center-d1.icon.n-tasks\endcsname{11400.0}
\expandafter\def\csname ppt@rev.desc.d7.uiv.tau-T-by-type.p1.center-d1.icon.n-tasks\endcsname{$+$11400.0}
\expandafter\def\csname val@rev.desc.d7.uiv.tau-T-by-type.p1.center-d1.icon.tau.ci90.0\endcsname{-0.088}
\expandafter\def\csname cnt@rev.desc.d7.uiv.tau-T-by-type.p1.center-d1.icon.tau.ci90.0\endcsname{0}
\expandafter\def\csname pct@rev.desc.d7.uiv.tau-T-by-type.p1.center-d1.icon.tau.ci90.0\endcsname{$-$8.8}
\expandafter\def\csname ppt@rev.desc.d7.uiv.tau-T-by-type.p1.center-d1.icon.tau.ci90.0\endcsname{$-$8.8}
\expandafter\def\csname val@rev.desc.d7.uiv.tau-T-by-type.p1.center-d1.icon.tau.ci90.1\endcsname{0.012}
\expandafter\def\csname cnt@rev.desc.d7.uiv.tau-T-by-type.p1.center-d1.icon.tau.ci90.1\endcsname{0}
\expandafter\def\csname pct@rev.desc.d7.uiv.tau-T-by-type.p1.center-d1.icon.tau.ci90.1\endcsname{1.2}
\expandafter\def\csname ppt@rev.desc.d7.uiv.tau-T-by-type.p1.center-d1.icon.tau.ci90.1\endcsname{$+$1.2}
\expandafter\def\csname val@rev.desc.d7.uiv.tau-T-by-type.p1.center-d1.icon.tau.ci95-excludes-0\endcsname{0}
\expandafter\def\csname cnt@rev.desc.d7.uiv.tau-T-by-type.p1.center-d1.icon.tau.ci95-excludes-0\endcsname{0}
\expandafter\def\csname pct@rev.desc.d7.uiv.tau-T-by-type.p1.center-d1.icon.tau.ci95-excludes-0\endcsname{0.0}
\expandafter\def\csname ppt@rev.desc.d7.uiv.tau-T-by-type.p1.center-d1.icon.tau.ci95-excludes-0\endcsname{0.0}
\expandafter\def\csname val@rev.desc.d7.uiv.tau-T-by-type.p1.center-d1.icon.tau.ci95.0\endcsname{-0.099}
\expandafter\def\csname cnt@rev.desc.d7.uiv.tau-T-by-type.p1.center-d1.icon.tau.ci95.0\endcsname{0}
\expandafter\def\csname pct@rev.desc.d7.uiv.tau-T-by-type.p1.center-d1.icon.tau.ci95.0\endcsname{$-$9.9}
\expandafter\def\csname ppt@rev.desc.d7.uiv.tau-T-by-type.p1.center-d1.icon.tau.ci95.0\endcsname{$-$9.9}
\expandafter\def\csname val@rev.desc.d7.uiv.tau-T-by-type.p1.center-d1.icon.tau.ci95.1\endcsname{0.017}
\expandafter\def\csname cnt@rev.desc.d7.uiv.tau-T-by-type.p1.center-d1.icon.tau.ci95.1\endcsname{0}
\expandafter\def\csname pct@rev.desc.d7.uiv.tau-T-by-type.p1.center-d1.icon.tau.ci95.1\endcsname{1.7}
\expandafter\def\csname ppt@rev.desc.d7.uiv.tau-T-by-type.p1.center-d1.icon.tau.ci95.1\endcsname{$+$1.7}
\expandafter\def\csname val@rev.desc.d7.uiv.tau-T-by-type.p1.center-d1.icon.tau.mean\endcsname{-0.036}
\expandafter\def\csname cnt@rev.desc.d7.uiv.tau-T-by-type.p1.center-d1.icon.tau.mean\endcsname{0}
\expandafter\def\csname pct@rev.desc.d7.uiv.tau-T-by-type.p1.center-d1.icon.tau.mean\endcsname{$-$3.6}
\expandafter\def\csname ppt@rev.desc.d7.uiv.tau-T-by-type.p1.center-d1.icon.tau.mean\endcsname{$-$3.6}
\expandafter\def\csname val@rev.desc.d7.uiv.tau-T-by-type.p1.center-d1.icon.tau.n-apps\endcsname{23}
\expandafter\def\csname cnt@rev.desc.d7.uiv.tau-T-by-type.p1.center-d1.icon.tau.n-apps\endcsname{23}
\expandafter\def\csname pct@rev.desc.d7.uiv.tau-T-by-type.p1.center-d1.icon.tau.n-apps\endcsname{2300.0}
\expandafter\def\csname ppt@rev.desc.d7.uiv.tau-T-by-type.p1.center-d1.icon.tau.n-apps\endcsname{$+$2300.0}
\expandafter\def\csname val@rev.desc.d7.uiv.tau-T-by-type.p1.center-d1.icon.tau.tost1.ci90-inside\endcsname{0}
\expandafter\def\csname cnt@rev.desc.d7.uiv.tau-T-by-type.p1.center-d1.icon.tau.tost1.ci90-inside\endcsname{0}
\expandafter\def\csname pct@rev.desc.d7.uiv.tau-T-by-type.p1.center-d1.icon.tau.tost1.ci90-inside\endcsname{0.0}
\expandafter\def\csname ppt@rev.desc.d7.uiv.tau-T-by-type.p1.center-d1.icon.tau.tost1.ci90-inside\endcsname{0.0}
\expandafter\def\csname val@rev.desc.d7.uiv.tau-T-by-type.p1.center-d1.icon.tau.tost1.equivalent\endcsname{0}
\expandafter\def\csname cnt@rev.desc.d7.uiv.tau-T-by-type.p1.center-d1.icon.tau.tost1.equivalent\endcsname{0}
\expandafter\def\csname pct@rev.desc.d7.uiv.tau-T-by-type.p1.center-d1.icon.tau.tost1.equivalent\endcsname{0.0}
\expandafter\def\csname ppt@rev.desc.d7.uiv.tau-T-by-type.p1.center-d1.icon.tau.tost1.equivalent\endcsname{0.0}
\expandafter\def\csname val@rev.desc.d7.uiv.tau-T-by-type.p1.center-d1.icon.tau.tost1.p\endcsname{0.808}
\expandafter\def\csname cnt@rev.desc.d7.uiv.tau-T-by-type.p1.center-d1.icon.tau.tost1.p\endcsname{1}
\expandafter\def\csname pct@rev.desc.d7.uiv.tau-T-by-type.p1.center-d1.icon.tau.tost1.p\endcsname{80.8}
\expandafter\def\csname ppt@rev.desc.d7.uiv.tau-T-by-type.p1.center-d1.icon.tau.tost1.p\endcsname{$+$80.8}
\expandafter\def\csname val@rev.desc.d7.uiv.tau-T-by-type.p1.center-d1.icon.tau.tost5.ci90-inside\endcsname{0}
\expandafter\def\csname cnt@rev.desc.d7.uiv.tau-T-by-type.p1.center-d1.icon.tau.tost5.ci90-inside\endcsname{0}
\expandafter\def\csname pct@rev.desc.d7.uiv.tau-T-by-type.p1.center-d1.icon.tau.tost5.ci90-inside\endcsname{0.0}
\expandafter\def\csname ppt@rev.desc.d7.uiv.tau-T-by-type.p1.center-d1.icon.tau.tost5.ci90-inside\endcsname{0.0}
\expandafter\def\csname val@rev.desc.d7.uiv.tau-T-by-type.p1.center-d1.icon.tau.tost5.equivalent\endcsname{0}
\expandafter\def\csname cnt@rev.desc.d7.uiv.tau-T-by-type.p1.center-d1.icon.tau.tost5.equivalent\endcsname{0}
\expandafter\def\csname pct@rev.desc.d7.uiv.tau-T-by-type.p1.center-d1.icon.tau.tost5.equivalent\endcsname{0.0}
\expandafter\def\csname ppt@rev.desc.d7.uiv.tau-T-by-type.p1.center-d1.icon.tau.tost5.equivalent\endcsname{0.0}
\expandafter\def\csname val@rev.desc.d7.uiv.tau-T-by-type.p1.center-d1.icon.tau.tost5.p\endcsname{0.318}
\expandafter\def\csname cnt@rev.desc.d7.uiv.tau-T-by-type.p1.center-d1.icon.tau.tost5.p\endcsname{0}
\expandafter\def\csname pct@rev.desc.d7.uiv.tau-T-by-type.p1.center-d1.icon.tau.tost5.p\endcsname{31.8}
\expandafter\def\csname ppt@rev.desc.d7.uiv.tau-T-by-type.p1.center-d1.icon.tau.tost5.p\endcsname{$+$31.8}
\expandafter\def\csname val@rev.desc.d7.uiv.tau-T-by-type.p1.center-d1.text.acc-F\endcsname{0.767}
\expandafter\def\csname cnt@rev.desc.d7.uiv.tau-T-by-type.p1.center-d1.text.acc-F\endcsname{1}
\expandafter\def\csname pct@rev.desc.d7.uiv.tau-T-by-type.p1.center-d1.text.acc-F\endcsname{76.7}
\expandafter\def\csname ppt@rev.desc.d7.uiv.tau-T-by-type.p1.center-d1.text.acc-F\endcsname{$+$76.7}
\expandafter\def\csname val@rev.desc.d7.uiv.tau-T-by-type.p1.center-d1.text.acc-direct\endcsname{0.867}
\expandafter\def\csname cnt@rev.desc.d7.uiv.tau-T-by-type.p1.center-d1.text.acc-direct\endcsname{1}
\expandafter\def\csname pct@rev.desc.d7.uiv.tau-T-by-type.p1.center-d1.text.acc-direct\endcsname{86.7}
\expandafter\def\csname ppt@rev.desc.d7.uiv.tau-T-by-type.p1.center-d1.text.acc-direct\endcsname{$+$86.7}
\expandafter\def\csname val@rev.desc.d7.uiv.tau-T-by-type.p1.center-d1.text.n-apps\endcsname{15}
\expandafter\def\csname cnt@rev.desc.d7.uiv.tau-T-by-type.p1.center-d1.text.n-apps\endcsname{15}
\expandafter\def\csname pct@rev.desc.d7.uiv.tau-T-by-type.p1.center-d1.text.n-apps\endcsname{1500.0}
\expandafter\def\csname ppt@rev.desc.d7.uiv.tau-T-by-type.p1.center-d1.text.n-apps\endcsname{$+$1500.0}
\expandafter\def\csname val@rev.desc.d7.uiv.tau-T-by-type.p1.center-d1.text.n-tasks\endcsname{30}
\expandafter\def\csname cnt@rev.desc.d7.uiv.tau-T-by-type.p1.center-d1.text.n-tasks\endcsname{30}
\expandafter\def\csname pct@rev.desc.d7.uiv.tau-T-by-type.p1.center-d1.text.n-tasks\endcsname{3000.0}
\expandafter\def\csname ppt@rev.desc.d7.uiv.tau-T-by-type.p1.center-d1.text.n-tasks\endcsname{$+$3000.0}
\expandafter\def\csname val@rev.desc.d7.uiv.tau-T-by-type.p1.center-d1.text.tau.ci90.0\endcsname{-0.217}
\expandafter\def\csname cnt@rev.desc.d7.uiv.tau-T-by-type.p1.center-d1.text.tau.ci90.0\endcsname{0}
\expandafter\def\csname pct@rev.desc.d7.uiv.tau-T-by-type.p1.center-d1.text.tau.ci90.0\endcsname{$-$21.7}
\expandafter\def\csname ppt@rev.desc.d7.uiv.tau-T-by-type.p1.center-d1.text.tau.ci90.0\endcsname{$-$21.7}
\expandafter\def\csname val@rev.desc.d7.uiv.tau-T-by-type.p1.center-d1.text.tau.ci90.1\endcsname{-0.011}
\expandafter\def\csname cnt@rev.desc.d7.uiv.tau-T-by-type.p1.center-d1.text.tau.ci90.1\endcsname{0}
\expandafter\def\csname pct@rev.desc.d7.uiv.tau-T-by-type.p1.center-d1.text.tau.ci90.1\endcsname{$-$1.1}
\expandafter\def\csname ppt@rev.desc.d7.uiv.tau-T-by-type.p1.center-d1.text.tau.ci90.1\endcsname{$-$1.1}
\expandafter\def\csname val@rev.desc.d7.uiv.tau-T-by-type.p1.center-d1.text.tau.ci95-excludes-0\endcsname{0}
\expandafter\def\csname cnt@rev.desc.d7.uiv.tau-T-by-type.p1.center-d1.text.tau.ci95-excludes-0\endcsname{0}
\expandafter\def\csname pct@rev.desc.d7.uiv.tau-T-by-type.p1.center-d1.text.tau.ci95-excludes-0\endcsname{0.0}
\expandafter\def\csname ppt@rev.desc.d7.uiv.tau-T-by-type.p1.center-d1.text.tau.ci95-excludes-0\endcsname{0.0}
\expandafter\def\csname val@rev.desc.d7.uiv.tau-T-by-type.p1.center-d1.text.tau.ci95.0\endcsname{-0.239}
\expandafter\def\csname cnt@rev.desc.d7.uiv.tau-T-by-type.p1.center-d1.text.tau.ci95.0\endcsname{0}
\expandafter\def\csname pct@rev.desc.d7.uiv.tau-T-by-type.p1.center-d1.text.tau.ci95.0\endcsname{$-$23.9}
\expandafter\def\csname ppt@rev.desc.d7.uiv.tau-T-by-type.p1.center-d1.text.tau.ci95.0\endcsname{$-$23.9}
\expandafter\def\csname val@rev.desc.d7.uiv.tau-T-by-type.p1.center-d1.text.tau.ci95.1\endcsname{0.000}
\expandafter\def\csname cnt@rev.desc.d7.uiv.tau-T-by-type.p1.center-d1.text.tau.ci95.1\endcsname{0}
\expandafter\def\csname pct@rev.desc.d7.uiv.tau-T-by-type.p1.center-d1.text.tau.ci95.1\endcsname{0.0}
\expandafter\def\csname ppt@rev.desc.d7.uiv.tau-T-by-type.p1.center-d1.text.tau.ci95.1\endcsname{0.0}
\expandafter\def\csname val@rev.desc.d7.uiv.tau-T-by-type.p1.center-d1.text.tau.mean\endcsname{-0.094}
\expandafter\def\csname cnt@rev.desc.d7.uiv.tau-T-by-type.p1.center-d1.text.tau.mean\endcsname{0}
\expandafter\def\csname pct@rev.desc.d7.uiv.tau-T-by-type.p1.center-d1.text.tau.mean\endcsname{$-$9.4}
\expandafter\def\csname ppt@rev.desc.d7.uiv.tau-T-by-type.p1.center-d1.text.tau.mean\endcsname{$-$9.4}
\expandafter\def\csname val@rev.desc.d7.uiv.tau-T-by-type.p1.center-d1.text.tau.n-apps\endcsname{15}
\expandafter\def\csname cnt@rev.desc.d7.uiv.tau-T-by-type.p1.center-d1.text.tau.n-apps\endcsname{15}
\expandafter\def\csname pct@rev.desc.d7.uiv.tau-T-by-type.p1.center-d1.text.tau.n-apps\endcsname{1500.0}
\expandafter\def\csname ppt@rev.desc.d7.uiv.tau-T-by-type.p1.center-d1.text.tau.n-apps\endcsname{$+$1500.0}
\expandafter\def\csname val@rev.desc.d7.uiv.tau-T-by-type.p1.center-d1.text.tau.tost1.ci90-inside\endcsname{0}
\expandafter\def\csname cnt@rev.desc.d7.uiv.tau-T-by-type.p1.center-d1.text.tau.tost1.ci90-inside\endcsname{0}
\expandafter\def\csname pct@rev.desc.d7.uiv.tau-T-by-type.p1.center-d1.text.tau.tost1.ci90-inside\endcsname{0.0}
\expandafter\def\csname ppt@rev.desc.d7.uiv.tau-T-by-type.p1.center-d1.text.tau.tost1.ci90-inside\endcsname{0.0}
\expandafter\def\csname val@rev.desc.d7.uiv.tau-T-by-type.p1.center-d1.text.tau.tost1.equivalent\endcsname{0}
\expandafter\def\csname cnt@rev.desc.d7.uiv.tau-T-by-type.p1.center-d1.text.tau.tost1.equivalent\endcsname{0}
\expandafter\def\csname pct@rev.desc.d7.uiv.tau-T-by-type.p1.center-d1.text.tau.tost1.equivalent\endcsname{0.0}
\expandafter\def\csname ppt@rev.desc.d7.uiv.tau-T-by-type.p1.center-d1.text.tau.tost1.equivalent\endcsname{0.0}
\expandafter\def\csname val@rev.desc.d7.uiv.tau-T-by-type.p1.center-d1.text.tau.tost1.p\endcsname{0.969}
\expandafter\def\csname cnt@rev.desc.d7.uiv.tau-T-by-type.p1.center-d1.text.tau.tost1.p\endcsname{1}
\expandafter\def\csname pct@rev.desc.d7.uiv.tau-T-by-type.p1.center-d1.text.tau.tost1.p\endcsname{96.9}
\expandafter\def\csname ppt@rev.desc.d7.uiv.tau-T-by-type.p1.center-d1.text.tau.tost1.p\endcsname{$+$96.9}
\expandafter\def\csname val@rev.desc.d7.uiv.tau-T-by-type.p1.center-d1.text.tau.tost5.ci90-inside\endcsname{0}
\expandafter\def\csname cnt@rev.desc.d7.uiv.tau-T-by-type.p1.center-d1.text.tau.tost5.ci90-inside\endcsname{0}
\expandafter\def\csname pct@rev.desc.d7.uiv.tau-T-by-type.p1.center-d1.text.tau.tost5.ci90-inside\endcsname{0.0}
\expandafter\def\csname ppt@rev.desc.d7.uiv.tau-T-by-type.p1.center-d1.text.tau.tost5.ci90-inside\endcsname{0.0}
\expandafter\def\csname val@rev.desc.d7.uiv.tau-T-by-type.p1.center-d1.text.tau.tost5.equivalent\endcsname{0}
\expandafter\def\csname cnt@rev.desc.d7.uiv.tau-T-by-type.p1.center-d1.text.tau.tost5.equivalent\endcsname{0}
\expandafter\def\csname pct@rev.desc.d7.uiv.tau-T-by-type.p1.center-d1.text.tau.tost5.equivalent\endcsname{0.0}
\expandafter\def\csname ppt@rev.desc.d7.uiv.tau-T-by-type.p1.center-d1.text.tau.tost5.equivalent\endcsname{0.0}
\expandafter\def\csname val@rev.desc.d7.uiv.tau-T-by-type.p1.center-d1.text.tau.tost5.p\endcsname{0.698}
\expandafter\def\csname cnt@rev.desc.d7.uiv.tau-T-by-type.p1.center-d1.text.tau.tost5.p\endcsname{1}
\expandafter\def\csname pct@rev.desc.d7.uiv.tau-T-by-type.p1.center-d1.text.tau.tost5.p\endcsname{69.8}
\expandafter\def\csname ppt@rev.desc.d7.uiv.tau-T-by-type.p1.center-d1.text.tau.tost5.p\endcsname{$+$69.8}
\expandafter\def\csname val@rev.desc.d7.uiv.tau-T-by-type.p1.context-d1.icon-minus-text-on-apps-with-both.ci90.0\endcsname{-0.014}
\expandafter\def\csname cnt@rev.desc.d7.uiv.tau-T-by-type.p1.context-d1.icon-minus-text-on-apps-with-both.ci90.0\endcsname{0}
\expandafter\def\csname pct@rev.desc.d7.uiv.tau-T-by-type.p1.context-d1.icon-minus-text-on-apps-with-both.ci90.0\endcsname{$-$1.4}
\expandafter\def\csname ppt@rev.desc.d7.uiv.tau-T-by-type.p1.context-d1.icon-minus-text-on-apps-with-both.ci90.0\endcsname{$-$1.4}
\expandafter\def\csname val@rev.desc.d7.uiv.tau-T-by-type.p1.context-d1.icon-minus-text-on-apps-with-both.ci90.1\endcsname{0.146}
\expandafter\def\csname cnt@rev.desc.d7.uiv.tau-T-by-type.p1.context-d1.icon-minus-text-on-apps-with-both.ci90.1\endcsname{0}
\expandafter\def\csname pct@rev.desc.d7.uiv.tau-T-by-type.p1.context-d1.icon-minus-text-on-apps-with-both.ci90.1\endcsname{14.6}
\expandafter\def\csname ppt@rev.desc.d7.uiv.tau-T-by-type.p1.context-d1.icon-minus-text-on-apps-with-both.ci90.1\endcsname{$+$14.6}
\expandafter\def\csname val@rev.desc.d7.uiv.tau-T-by-type.p1.context-d1.icon-minus-text-on-apps-with-both.ci95-excludes-0\endcsname{0}
\expandafter\def\csname cnt@rev.desc.d7.uiv.tau-T-by-type.p1.context-d1.icon-minus-text-on-apps-with-both.ci95-excludes-0\endcsname{0}
\expandafter\def\csname pct@rev.desc.d7.uiv.tau-T-by-type.p1.context-d1.icon-minus-text-on-apps-with-both.ci95-excludes-0\endcsname{0.0}
\expandafter\def\csname ppt@rev.desc.d7.uiv.tau-T-by-type.p1.context-d1.icon-minus-text-on-apps-with-both.ci95-excludes-0\endcsname{0.0}
\expandafter\def\csname val@rev.desc.d7.uiv.tau-T-by-type.p1.context-d1.icon-minus-text-on-apps-with-both.ci95.0\endcsname{-0.014}
\expandafter\def\csname cnt@rev.desc.d7.uiv.tau-T-by-type.p1.context-d1.icon-minus-text-on-apps-with-both.ci95.0\endcsname{0}
\expandafter\def\csname pct@rev.desc.d7.uiv.tau-T-by-type.p1.context-d1.icon-minus-text-on-apps-with-both.ci95.0\endcsname{$-$1.4}
\expandafter\def\csname ppt@rev.desc.d7.uiv.tau-T-by-type.p1.context-d1.icon-minus-text-on-apps-with-both.ci95.0\endcsname{$-$1.4}
\expandafter\def\csname val@rev.desc.d7.uiv.tau-T-by-type.p1.context-d1.icon-minus-text-on-apps-with-both.ci95.1\endcsname{0.164}
\expandafter\def\csname cnt@rev.desc.d7.uiv.tau-T-by-type.p1.context-d1.icon-minus-text-on-apps-with-both.ci95.1\endcsname{0}
\expandafter\def\csname pct@rev.desc.d7.uiv.tau-T-by-type.p1.context-d1.icon-minus-text-on-apps-with-both.ci95.1\endcsname{16.4}
\expandafter\def\csname ppt@rev.desc.d7.uiv.tau-T-by-type.p1.context-d1.icon-minus-text-on-apps-with-both.ci95.1\endcsname{$+$16.4}
\expandafter\def\csname val@rev.desc.d7.uiv.tau-T-by-type.p1.context-d1.icon-minus-text-on-apps-with-both.mean\endcsname{0.061}
\expandafter\def\csname cnt@rev.desc.d7.uiv.tau-T-by-type.p1.context-d1.icon-minus-text-on-apps-with-both.mean\endcsname{0}
\expandafter\def\csname pct@rev.desc.d7.uiv.tau-T-by-type.p1.context-d1.icon-minus-text-on-apps-with-both.mean\endcsname{6.1}
\expandafter\def\csname ppt@rev.desc.d7.uiv.tau-T-by-type.p1.context-d1.icon-minus-text-on-apps-with-both.mean\endcsname{$+$6.1}
\expandafter\def\csname val@rev.desc.d7.uiv.tau-T-by-type.p1.context-d1.icon-minus-text-on-apps-with-both.n-apps\endcsname{14}
\expandafter\def\csname cnt@rev.desc.d7.uiv.tau-T-by-type.p1.context-d1.icon-minus-text-on-apps-with-both.n-apps\endcsname{14}
\expandafter\def\csname pct@rev.desc.d7.uiv.tau-T-by-type.p1.context-d1.icon-minus-text-on-apps-with-both.n-apps\endcsname{1400.0}
\expandafter\def\csname ppt@rev.desc.d7.uiv.tau-T-by-type.p1.context-d1.icon-minus-text-on-apps-with-both.n-apps\endcsname{$+$1400.0}
\expandafter\def\csname val@rev.desc.d7.uiv.tau-T-by-type.p1.context-d1.icon-minus-text-on-apps-with-both.tost1.ci90-inside\endcsname{0}
\expandafter\def\csname cnt@rev.desc.d7.uiv.tau-T-by-type.p1.context-d1.icon-minus-text-on-apps-with-both.tost1.ci90-inside\endcsname{0}
\expandafter\def\csname pct@rev.desc.d7.uiv.tau-T-by-type.p1.context-d1.icon-minus-text-on-apps-with-both.tost1.ci90-inside\endcsname{0.0}
\expandafter\def\csname ppt@rev.desc.d7.uiv.tau-T-by-type.p1.context-d1.icon-minus-text-on-apps-with-both.tost1.ci90-inside\endcsname{0.0}
\expandafter\def\csname val@rev.desc.d7.uiv.tau-T-by-type.p1.context-d1.icon-minus-text-on-apps-with-both.tost1.equivalent\endcsname{0}
\expandafter\def\csname cnt@rev.desc.d7.uiv.tau-T-by-type.p1.context-d1.icon-minus-text-on-apps-with-both.tost1.equivalent\endcsname{0}
\expandafter\def\csname pct@rev.desc.d7.uiv.tau-T-by-type.p1.context-d1.icon-minus-text-on-apps-with-both.tost1.equivalent\endcsname{0.0}
\expandafter\def\csname ppt@rev.desc.d7.uiv.tau-T-by-type.p1.context-d1.icon-minus-text-on-apps-with-both.tost1.equivalent\endcsname{0.0}
\expandafter\def\csname val@rev.desc.d7.uiv.tau-T-by-type.p1.context-d1.icon-minus-text-on-apps-with-both.tost1.p\endcsname{0.860}
\expandafter\def\csname cnt@rev.desc.d7.uiv.tau-T-by-type.p1.context-d1.icon-minus-text-on-apps-with-both.tost1.p\endcsname{1}
\expandafter\def\csname pct@rev.desc.d7.uiv.tau-T-by-type.p1.context-d1.icon-minus-text-on-apps-with-both.tost1.p\endcsname{86.0}
\expandafter\def\csname ppt@rev.desc.d7.uiv.tau-T-by-type.p1.context-d1.icon-minus-text-on-apps-with-both.tost1.p\endcsname{$+$86.0}
\expandafter\def\csname val@rev.desc.d7.uiv.tau-T-by-type.p1.context-d1.icon-minus-text-on-apps-with-both.tost5.ci90-inside\endcsname{0}
\expandafter\def\csname cnt@rev.desc.d7.uiv.tau-T-by-type.p1.context-d1.icon-minus-text-on-apps-with-both.tost5.ci90-inside\endcsname{0}
\expandafter\def\csname pct@rev.desc.d7.uiv.tau-T-by-type.p1.context-d1.icon-minus-text-on-apps-with-both.tost5.ci90-inside\endcsname{0.0}
\expandafter\def\csname ppt@rev.desc.d7.uiv.tau-T-by-type.p1.context-d1.icon-minus-text-on-apps-with-both.tost5.ci90-inside\endcsname{0.0}
\expandafter\def\csname val@rev.desc.d7.uiv.tau-T-by-type.p1.context-d1.icon-minus-text-on-apps-with-both.tost5.equivalent\endcsname{0}
\expandafter\def\csname cnt@rev.desc.d7.uiv.tau-T-by-type.p1.context-d1.icon-minus-text-on-apps-with-both.tost5.equivalent\endcsname{0}
\expandafter\def\csname pct@rev.desc.d7.uiv.tau-T-by-type.p1.context-d1.icon-minus-text-on-apps-with-both.tost5.equivalent\endcsname{0.0}
\expandafter\def\csname ppt@rev.desc.d7.uiv.tau-T-by-type.p1.context-d1.icon-minus-text-on-apps-with-both.tost5.equivalent\endcsname{0.0}
\expandafter\def\csname val@rev.desc.d7.uiv.tau-T-by-type.p1.context-d1.icon-minus-text-on-apps-with-both.tost5.p\endcsname{0.535}
\expandafter\def\csname cnt@rev.desc.d7.uiv.tau-T-by-type.p1.context-d1.icon-minus-text-on-apps-with-both.tost5.p\endcsname{1}
\expandafter\def\csname pct@rev.desc.d7.uiv.tau-T-by-type.p1.context-d1.icon-minus-text-on-apps-with-both.tost5.p\endcsname{53.5}
\expandafter\def\csname ppt@rev.desc.d7.uiv.tau-T-by-type.p1.context-d1.icon-minus-text-on-apps-with-both.tost5.p\endcsname{$+$53.5}
\expandafter\def\csname val@rev.desc.d7.uiv.tau-T-by-type.p1.context-d1.icon.acc-F\endcsname{0.781}
\expandafter\def\csname cnt@rev.desc.d7.uiv.tau-T-by-type.p1.context-d1.icon.acc-F\endcsname{1}
\expandafter\def\csname pct@rev.desc.d7.uiv.tau-T-by-type.p1.context-d1.icon.acc-F\endcsname{78.1}
\expandafter\def\csname ppt@rev.desc.d7.uiv.tau-T-by-type.p1.context-d1.icon.acc-F\endcsname{$+$78.1}
\expandafter\def\csname val@rev.desc.d7.uiv.tau-T-by-type.p1.context-d1.icon.acc-direct\endcsname{0.825}
\expandafter\def\csname cnt@rev.desc.d7.uiv.tau-T-by-type.p1.context-d1.icon.acc-direct\endcsname{1}
\expandafter\def\csname pct@rev.desc.d7.uiv.tau-T-by-type.p1.context-d1.icon.acc-direct\endcsname{82.5}
\expandafter\def\csname ppt@rev.desc.d7.uiv.tau-T-by-type.p1.context-d1.icon.acc-direct\endcsname{$+$82.5}
\expandafter\def\csname val@rev.desc.d7.uiv.tau-T-by-type.p1.context-d1.icon.n-apps\endcsname{23}
\expandafter\def\csname cnt@rev.desc.d7.uiv.tau-T-by-type.p1.context-d1.icon.n-apps\endcsname{23}
\expandafter\def\csname pct@rev.desc.d7.uiv.tau-T-by-type.p1.context-d1.icon.n-apps\endcsname{2300.0}
\expandafter\def\csname ppt@rev.desc.d7.uiv.tau-T-by-type.p1.context-d1.icon.n-apps\endcsname{$+$2300.0}
\expandafter\def\csname val@rev.desc.d7.uiv.tau-T-by-type.p1.context-d1.icon.n-tasks\endcsname{114}
\expandafter\def\csname cnt@rev.desc.d7.uiv.tau-T-by-type.p1.context-d1.icon.n-tasks\endcsname{114}
\expandafter\def\csname pct@rev.desc.d7.uiv.tau-T-by-type.p1.context-d1.icon.n-tasks\endcsname{11400.0}
\expandafter\def\csname ppt@rev.desc.d7.uiv.tau-T-by-type.p1.context-d1.icon.n-tasks\endcsname{$+$11400.0}
\expandafter\def\csname val@rev.desc.d7.uiv.tau-T-by-type.p1.context-d1.icon.tau.ci90.0\endcsname{-0.084}
\expandafter\def\csname cnt@rev.desc.d7.uiv.tau-T-by-type.p1.context-d1.icon.tau.ci90.0\endcsname{0}
\expandafter\def\csname pct@rev.desc.d7.uiv.tau-T-by-type.p1.context-d1.icon.tau.ci90.0\endcsname{$-$8.4}
\expandafter\def\csname ppt@rev.desc.d7.uiv.tau-T-by-type.p1.context-d1.icon.tau.ci90.0\endcsname{$-$8.4}
\expandafter\def\csname val@rev.desc.d7.uiv.tau-T-by-type.p1.context-d1.icon.tau.ci90.1\endcsname{-0.000}
\expandafter\def\csname cnt@rev.desc.d7.uiv.tau-T-by-type.p1.context-d1.icon.tau.ci90.1\endcsname{0}
\expandafter\def\csname pct@rev.desc.d7.uiv.tau-T-by-type.p1.context-d1.icon.tau.ci90.1\endcsname{0.0}
\expandafter\def\csname ppt@rev.desc.d7.uiv.tau-T-by-type.p1.context-d1.icon.tau.ci90.1\endcsname{0.0}
\expandafter\def\csname val@rev.desc.d7.uiv.tau-T-by-type.p1.context-d1.icon.tau.ci95-excludes-0\endcsname{0}
\expandafter\def\csname cnt@rev.desc.d7.uiv.tau-T-by-type.p1.context-d1.icon.tau.ci95-excludes-0\endcsname{0}
\expandafter\def\csname pct@rev.desc.d7.uiv.tau-T-by-type.p1.context-d1.icon.tau.ci95-excludes-0\endcsname{0.0}
\expandafter\def\csname ppt@rev.desc.d7.uiv.tau-T-by-type.p1.context-d1.icon.tau.ci95-excludes-0\endcsname{0.0}
\expandafter\def\csname val@rev.desc.d7.uiv.tau-T-by-type.p1.context-d1.icon.tau.ci95.0\endcsname{-0.094}
\expandafter\def\csname cnt@rev.desc.d7.uiv.tau-T-by-type.p1.context-d1.icon.tau.ci95.0\endcsname{0}
\expandafter\def\csname pct@rev.desc.d7.uiv.tau-T-by-type.p1.context-d1.icon.tau.ci95.0\endcsname{$-$9.4}
\expandafter\def\csname ppt@rev.desc.d7.uiv.tau-T-by-type.p1.context-d1.icon.tau.ci95.0\endcsname{$-$9.4}
\expandafter\def\csname val@rev.desc.d7.uiv.tau-T-by-type.p1.context-d1.icon.tau.ci95.1\endcsname{0.004}
\expandafter\def\csname cnt@rev.desc.d7.uiv.tau-T-by-type.p1.context-d1.icon.tau.ci95.1\endcsname{0}
\expandafter\def\csname pct@rev.desc.d7.uiv.tau-T-by-type.p1.context-d1.icon.tau.ci95.1\endcsname{0.4}
\expandafter\def\csname ppt@rev.desc.d7.uiv.tau-T-by-type.p1.context-d1.icon.tau.ci95.1\endcsname{$+$0.4}
\expandafter\def\csname val@rev.desc.d7.uiv.tau-T-by-type.p1.context-d1.icon.tau.mean\endcsname{-0.039}
\expandafter\def\csname cnt@rev.desc.d7.uiv.tau-T-by-type.p1.context-d1.icon.tau.mean\endcsname{0}
\expandafter\def\csname pct@rev.desc.d7.uiv.tau-T-by-type.p1.context-d1.icon.tau.mean\endcsname{$-$3.9}
\expandafter\def\csname ppt@rev.desc.d7.uiv.tau-T-by-type.p1.context-d1.icon.tau.mean\endcsname{$-$3.9}
\expandafter\def\csname val@rev.desc.d7.uiv.tau-T-by-type.p1.context-d1.icon.tau.n-apps\endcsname{23}
\expandafter\def\csname cnt@rev.desc.d7.uiv.tau-T-by-type.p1.context-d1.icon.tau.n-apps\endcsname{23}
\expandafter\def\csname pct@rev.desc.d7.uiv.tau-T-by-type.p1.context-d1.icon.tau.n-apps\endcsname{2300.0}
\expandafter\def\csname ppt@rev.desc.d7.uiv.tau-T-by-type.p1.context-d1.icon.tau.n-apps\endcsname{$+$2300.0}
\expandafter\def\csname val@rev.desc.d7.uiv.tau-T-by-type.p1.context-d1.icon.tau.tost1.ci90-inside\endcsname{0}
\expandafter\def\csname cnt@rev.desc.d7.uiv.tau-T-by-type.p1.context-d1.icon.tau.tost1.ci90-inside\endcsname{0}
\expandafter\def\csname pct@rev.desc.d7.uiv.tau-T-by-type.p1.context-d1.icon.tau.tost1.ci90-inside\endcsname{0.0}
\expandafter\def\csname ppt@rev.desc.d7.uiv.tau-T-by-type.p1.context-d1.icon.tau.tost1.ci90-inside\endcsname{0.0}
\expandafter\def\csname val@rev.desc.d7.uiv.tau-T-by-type.p1.context-d1.icon.tau.tost1.equivalent\endcsname{0}
\expandafter\def\csname cnt@rev.desc.d7.uiv.tau-T-by-type.p1.context-d1.icon.tau.tost1.equivalent\endcsname{0}
\expandafter\def\csname pct@rev.desc.d7.uiv.tau-T-by-type.p1.context-d1.icon.tau.tost1.equivalent\endcsname{0.0}
\expandafter\def\csname ppt@rev.desc.d7.uiv.tau-T-by-type.p1.context-d1.icon.tau.tost1.equivalent\endcsname{0.0}
\expandafter\def\csname val@rev.desc.d7.uiv.tau-T-by-type.p1.context-d1.icon.tau.tost1.p\endcsname{0.872}
\expandafter\def\csname cnt@rev.desc.d7.uiv.tau-T-by-type.p1.context-d1.icon.tau.tost1.p\endcsname{1}
\expandafter\def\csname pct@rev.desc.d7.uiv.tau-T-by-type.p1.context-d1.icon.tau.tost1.p\endcsname{87.2}
\expandafter\def\csname ppt@rev.desc.d7.uiv.tau-T-by-type.p1.context-d1.icon.tau.tost1.p\endcsname{$+$87.2}
\expandafter\def\csname val@rev.desc.d7.uiv.tau-T-by-type.p1.context-d1.icon.tau.tost5.ci90-inside\endcsname{0}
\expandafter\def\csname cnt@rev.desc.d7.uiv.tau-T-by-type.p1.context-d1.icon.tau.tost5.ci90-inside\endcsname{0}
\expandafter\def\csname pct@rev.desc.d7.uiv.tau-T-by-type.p1.context-d1.icon.tau.tost5.ci90-inside\endcsname{0.0}
\expandafter\def\csname ppt@rev.desc.d7.uiv.tau-T-by-type.p1.context-d1.icon.tau.tost5.ci90-inside\endcsname{0.0}
\expandafter\def\csname val@rev.desc.d7.uiv.tau-T-by-type.p1.context-d1.icon.tau.tost5.equivalent\endcsname{0}
\expandafter\def\csname cnt@rev.desc.d7.uiv.tau-T-by-type.p1.context-d1.icon.tau.tost5.equivalent\endcsname{0}
\expandafter\def\csname pct@rev.desc.d7.uiv.tau-T-by-type.p1.context-d1.icon.tau.tost5.equivalent\endcsname{0.0}
\expandafter\def\csname ppt@rev.desc.d7.uiv.tau-T-by-type.p1.context-d1.icon.tau.tost5.equivalent\endcsname{0.0}
\expandafter\def\csname val@rev.desc.d7.uiv.tau-T-by-type.p1.context-d1.icon.tau.tost5.p\endcsname{0.319}
\expandafter\def\csname cnt@rev.desc.d7.uiv.tau-T-by-type.p1.context-d1.icon.tau.tost5.p\endcsname{0}
\expandafter\def\csname pct@rev.desc.d7.uiv.tau-T-by-type.p1.context-d1.icon.tau.tost5.p\endcsname{31.9}
\expandafter\def\csname ppt@rev.desc.d7.uiv.tau-T-by-type.p1.context-d1.icon.tau.tost5.p\endcsname{$+$31.9}
\expandafter\def\csname val@rev.desc.d7.uiv.tau-T-by-type.p1.context-d1.text.acc-F\endcsname{0.767}
\expandafter\def\csname cnt@rev.desc.d7.uiv.tau-T-by-type.p1.context-d1.text.acc-F\endcsname{1}
\expandafter\def\csname pct@rev.desc.d7.uiv.tau-T-by-type.p1.context-d1.text.acc-F\endcsname{76.7}
\expandafter\def\csname ppt@rev.desc.d7.uiv.tau-T-by-type.p1.context-d1.text.acc-F\endcsname{$+$76.7}
\expandafter\def\csname val@rev.desc.d7.uiv.tau-T-by-type.p1.context-d1.text.acc-direct\endcsname{0.867}
\expandafter\def\csname cnt@rev.desc.d7.uiv.tau-T-by-type.p1.context-d1.text.acc-direct\endcsname{1}
\expandafter\def\csname pct@rev.desc.d7.uiv.tau-T-by-type.p1.context-d1.text.acc-direct\endcsname{86.7}
\expandafter\def\csname ppt@rev.desc.d7.uiv.tau-T-by-type.p1.context-d1.text.acc-direct\endcsname{$+$86.7}
\expandafter\def\csname val@rev.desc.d7.uiv.tau-T-by-type.p1.context-d1.text.n-apps\endcsname{15}
\expandafter\def\csname cnt@rev.desc.d7.uiv.tau-T-by-type.p1.context-d1.text.n-apps\endcsname{15}
\expandafter\def\csname pct@rev.desc.d7.uiv.tau-T-by-type.p1.context-d1.text.n-apps\endcsname{1500.0}
\expandafter\def\csname ppt@rev.desc.d7.uiv.tau-T-by-type.p1.context-d1.text.n-apps\endcsname{$+$1500.0}
\expandafter\def\csname val@rev.desc.d7.uiv.tau-T-by-type.p1.context-d1.text.n-tasks\endcsname{30}
\expandafter\def\csname cnt@rev.desc.d7.uiv.tau-T-by-type.p1.context-d1.text.n-tasks\endcsname{30}
\expandafter\def\csname pct@rev.desc.d7.uiv.tau-T-by-type.p1.context-d1.text.n-tasks\endcsname{3000.0}
\expandafter\def\csname ppt@rev.desc.d7.uiv.tau-T-by-type.p1.context-d1.text.n-tasks\endcsname{$+$3000.0}
\expandafter\def\csname val@rev.desc.d7.uiv.tau-T-by-type.p1.context-d1.text.tau.ci90.0\endcsname{-0.217}
\expandafter\def\csname cnt@rev.desc.d7.uiv.tau-T-by-type.p1.context-d1.text.tau.ci90.0\endcsname{0}
\expandafter\def\csname pct@rev.desc.d7.uiv.tau-T-by-type.p1.context-d1.text.tau.ci90.0\endcsname{$-$21.7}
\expandafter\def\csname ppt@rev.desc.d7.uiv.tau-T-by-type.p1.context-d1.text.tau.ci90.0\endcsname{$-$21.7}
\expandafter\def\csname val@rev.desc.d7.uiv.tau-T-by-type.p1.context-d1.text.tau.ci90.1\endcsname{-0.011}
\expandafter\def\csname cnt@rev.desc.d7.uiv.tau-T-by-type.p1.context-d1.text.tau.ci90.1\endcsname{0}
\expandafter\def\csname pct@rev.desc.d7.uiv.tau-T-by-type.p1.context-d1.text.tau.ci90.1\endcsname{$-$1.1}
\expandafter\def\csname ppt@rev.desc.d7.uiv.tau-T-by-type.p1.context-d1.text.tau.ci90.1\endcsname{$-$1.1}
\expandafter\def\csname val@rev.desc.d7.uiv.tau-T-by-type.p1.context-d1.text.tau.ci95-excludes-0\endcsname{0}
\expandafter\def\csname cnt@rev.desc.d7.uiv.tau-T-by-type.p1.context-d1.text.tau.ci95-excludes-0\endcsname{0}
\expandafter\def\csname pct@rev.desc.d7.uiv.tau-T-by-type.p1.context-d1.text.tau.ci95-excludes-0\endcsname{0.0}
\expandafter\def\csname ppt@rev.desc.d7.uiv.tau-T-by-type.p1.context-d1.text.tau.ci95-excludes-0\endcsname{0.0}
\expandafter\def\csname val@rev.desc.d7.uiv.tau-T-by-type.p1.context-d1.text.tau.ci95.0\endcsname{-0.239}
\expandafter\def\csname cnt@rev.desc.d7.uiv.tau-T-by-type.p1.context-d1.text.tau.ci95.0\endcsname{0}
\expandafter\def\csname pct@rev.desc.d7.uiv.tau-T-by-type.p1.context-d1.text.tau.ci95.0\endcsname{$-$23.9}
\expandafter\def\csname ppt@rev.desc.d7.uiv.tau-T-by-type.p1.context-d1.text.tau.ci95.0\endcsname{$-$23.9}
\expandafter\def\csname val@rev.desc.d7.uiv.tau-T-by-type.p1.context-d1.text.tau.ci95.1\endcsname{0.000}
\expandafter\def\csname cnt@rev.desc.d7.uiv.tau-T-by-type.p1.context-d1.text.tau.ci95.1\endcsname{0}
\expandafter\def\csname pct@rev.desc.d7.uiv.tau-T-by-type.p1.context-d1.text.tau.ci95.1\endcsname{0.0}
\expandafter\def\csname ppt@rev.desc.d7.uiv.tau-T-by-type.p1.context-d1.text.tau.ci95.1\endcsname{0.0}
\expandafter\def\csname val@rev.desc.d7.uiv.tau-T-by-type.p1.context-d1.text.tau.mean\endcsname{-0.094}
\expandafter\def\csname cnt@rev.desc.d7.uiv.tau-T-by-type.p1.context-d1.text.tau.mean\endcsname{0}
\expandafter\def\csname pct@rev.desc.d7.uiv.tau-T-by-type.p1.context-d1.text.tau.mean\endcsname{$-$9.4}
\expandafter\def\csname ppt@rev.desc.d7.uiv.tau-T-by-type.p1.context-d1.text.tau.mean\endcsname{$-$9.4}
\expandafter\def\csname val@rev.desc.d7.uiv.tau-T-by-type.p1.context-d1.text.tau.n-apps\endcsname{15}
\expandafter\def\csname cnt@rev.desc.d7.uiv.tau-T-by-type.p1.context-d1.text.tau.n-apps\endcsname{15}
\expandafter\def\csname pct@rev.desc.d7.uiv.tau-T-by-type.p1.context-d1.text.tau.n-apps\endcsname{1500.0}
\expandafter\def\csname ppt@rev.desc.d7.uiv.tau-T-by-type.p1.context-d1.text.tau.n-apps\endcsname{$+$1500.0}
\expandafter\def\csname val@rev.desc.d7.uiv.tau-T-by-type.p1.context-d1.text.tau.tost1.ci90-inside\endcsname{0}
\expandafter\def\csname cnt@rev.desc.d7.uiv.tau-T-by-type.p1.context-d1.text.tau.tost1.ci90-inside\endcsname{0}
\expandafter\def\csname pct@rev.desc.d7.uiv.tau-T-by-type.p1.context-d1.text.tau.tost1.ci90-inside\endcsname{0.0}
\expandafter\def\csname ppt@rev.desc.d7.uiv.tau-T-by-type.p1.context-d1.text.tau.tost1.ci90-inside\endcsname{0.0}
\expandafter\def\csname val@rev.desc.d7.uiv.tau-T-by-type.p1.context-d1.text.tau.tost1.equivalent\endcsname{0}
\expandafter\def\csname cnt@rev.desc.d7.uiv.tau-T-by-type.p1.context-d1.text.tau.tost1.equivalent\endcsname{0}
\expandafter\def\csname pct@rev.desc.d7.uiv.tau-T-by-type.p1.context-d1.text.tau.tost1.equivalent\endcsname{0.0}
\expandafter\def\csname ppt@rev.desc.d7.uiv.tau-T-by-type.p1.context-d1.text.tau.tost1.equivalent\endcsname{0.0}
\expandafter\def\csname val@rev.desc.d7.uiv.tau-T-by-type.p1.context-d1.text.tau.tost1.p\endcsname{0.964}
\expandafter\def\csname cnt@rev.desc.d7.uiv.tau-T-by-type.p1.context-d1.text.tau.tost1.p\endcsname{1}
\expandafter\def\csname pct@rev.desc.d7.uiv.tau-T-by-type.p1.context-d1.text.tau.tost1.p\endcsname{96.4}
\expandafter\def\csname ppt@rev.desc.d7.uiv.tau-T-by-type.p1.context-d1.text.tau.tost1.p\endcsname{$+$96.4}
\expandafter\def\csname val@rev.desc.d7.uiv.tau-T-by-type.p1.context-d1.text.tau.tost5.ci90-inside\endcsname{0}
\expandafter\def\csname cnt@rev.desc.d7.uiv.tau-T-by-type.p1.context-d1.text.tau.tost5.ci90-inside\endcsname{0}
\expandafter\def\csname pct@rev.desc.d7.uiv.tau-T-by-type.p1.context-d1.text.tau.tost5.ci90-inside\endcsname{0.0}
\expandafter\def\csname ppt@rev.desc.d7.uiv.tau-T-by-type.p1.context-d1.text.tau.tost5.ci90-inside\endcsname{0.0}
\expandafter\def\csname val@rev.desc.d7.uiv.tau-T-by-type.p1.context-d1.text.tau.tost5.equivalent\endcsname{0}
\expandafter\def\csname cnt@rev.desc.d7.uiv.tau-T-by-type.p1.context-d1.text.tau.tost5.equivalent\endcsname{0}
\expandafter\def\csname pct@rev.desc.d7.uiv.tau-T-by-type.p1.context-d1.text.tau.tost5.equivalent\endcsname{0.0}
\expandafter\def\csname ppt@rev.desc.d7.uiv.tau-T-by-type.p1.context-d1.text.tau.tost5.equivalent\endcsname{0.0}
\expandafter\def\csname val@rev.desc.d7.uiv.tau-T-by-type.p1.context-d1.text.tau.tost5.p\endcsname{0.706}
\expandafter\def\csname cnt@rev.desc.d7.uiv.tau-T-by-type.p1.context-d1.text.tau.tost5.p\endcsname{1}
\expandafter\def\csname pct@rev.desc.d7.uiv.tau-T-by-type.p1.context-d1.text.tau.tost5.p\endcsname{70.6}
\expandafter\def\csname ppt@rev.desc.d7.uiv.tau-T-by-type.p1.context-d1.text.tau.tost5.p\endcsname{$+$70.6}
\expandafter\def\csname val@rev.desc.d7.uiv.tau-T-by-type.p1.dual-d2.icon-minus-text-on-apps-with-both.ci90.0\endcsname{-0.118}
\expandafter\def\csname cnt@rev.desc.d7.uiv.tau-T-by-type.p1.dual-d2.icon-minus-text-on-apps-with-both.ci90.0\endcsname{0}
\expandafter\def\csname pct@rev.desc.d7.uiv.tau-T-by-type.p1.dual-d2.icon-minus-text-on-apps-with-both.ci90.0\endcsname{$-$11.8}
\expandafter\def\csname ppt@rev.desc.d7.uiv.tau-T-by-type.p1.dual-d2.icon-minus-text-on-apps-with-both.ci90.0\endcsname{$-$11.8}
\expandafter\def\csname val@rev.desc.d7.uiv.tau-T-by-type.p1.dual-d2.icon-minus-text-on-apps-with-both.ci90.1\endcsname{0.096}
\expandafter\def\csname cnt@rev.desc.d7.uiv.tau-T-by-type.p1.dual-d2.icon-minus-text-on-apps-with-both.ci90.1\endcsname{0}
\expandafter\def\csname pct@rev.desc.d7.uiv.tau-T-by-type.p1.dual-d2.icon-minus-text-on-apps-with-both.ci90.1\endcsname{9.6}
\expandafter\def\csname ppt@rev.desc.d7.uiv.tau-T-by-type.p1.dual-d2.icon-minus-text-on-apps-with-both.ci90.1\endcsname{$+$9.6}
\expandafter\def\csname val@rev.desc.d7.uiv.tau-T-by-type.p1.dual-d2.icon-minus-text-on-apps-with-both.ci95-excludes-0\endcsname{0}
\expandafter\def\csname cnt@rev.desc.d7.uiv.tau-T-by-type.p1.dual-d2.icon-minus-text-on-apps-with-both.ci95-excludes-0\endcsname{0}
\expandafter\def\csname pct@rev.desc.d7.uiv.tau-T-by-type.p1.dual-d2.icon-minus-text-on-apps-with-both.ci95-excludes-0\endcsname{0.0}
\expandafter\def\csname ppt@rev.desc.d7.uiv.tau-T-by-type.p1.dual-d2.icon-minus-text-on-apps-with-both.ci95-excludes-0\endcsname{0.0}
\expandafter\def\csname val@rev.desc.d7.uiv.tau-T-by-type.p1.dual-d2.icon-minus-text-on-apps-with-both.ci95.0\endcsname{-0.129}
\expandafter\def\csname cnt@rev.desc.d7.uiv.tau-T-by-type.p1.dual-d2.icon-minus-text-on-apps-with-both.ci95.0\endcsname{0}
\expandafter\def\csname pct@rev.desc.d7.uiv.tau-T-by-type.p1.dual-d2.icon-minus-text-on-apps-with-both.ci95.0\endcsname{$-$12.9}
\expandafter\def\csname ppt@rev.desc.d7.uiv.tau-T-by-type.p1.dual-d2.icon-minus-text-on-apps-with-both.ci95.0\endcsname{$-$12.9}
\expandafter\def\csname val@rev.desc.d7.uiv.tau-T-by-type.p1.dual-d2.icon-minus-text-on-apps-with-both.ci95.1\endcsname{0.121}
\expandafter\def\csname cnt@rev.desc.d7.uiv.tau-T-by-type.p1.dual-d2.icon-minus-text-on-apps-with-both.ci95.1\endcsname{0}
\expandafter\def\csname pct@rev.desc.d7.uiv.tau-T-by-type.p1.dual-d2.icon-minus-text-on-apps-with-both.ci95.1\endcsname{12.1}
\expandafter\def\csname ppt@rev.desc.d7.uiv.tau-T-by-type.p1.dual-d2.icon-minus-text-on-apps-with-both.ci95.1\endcsname{$+$12.1}
\expandafter\def\csname val@rev.desc.d7.uiv.tau-T-by-type.p1.dual-d2.icon-minus-text-on-apps-with-both.mean\endcsname{-0.021}
\expandafter\def\csname cnt@rev.desc.d7.uiv.tau-T-by-type.p1.dual-d2.icon-minus-text-on-apps-with-both.mean\endcsname{0}
\expandafter\def\csname pct@rev.desc.d7.uiv.tau-T-by-type.p1.dual-d2.icon-minus-text-on-apps-with-both.mean\endcsname{$-$2.1}
\expandafter\def\csname ppt@rev.desc.d7.uiv.tau-T-by-type.p1.dual-d2.icon-minus-text-on-apps-with-both.mean\endcsname{$-$2.1}
\expandafter\def\csname val@rev.desc.d7.uiv.tau-T-by-type.p1.dual-d2.icon-minus-text-on-apps-with-both.n-apps\endcsname{14}
\expandafter\def\csname cnt@rev.desc.d7.uiv.tau-T-by-type.p1.dual-d2.icon-minus-text-on-apps-with-both.n-apps\endcsname{14}
\expandafter\def\csname pct@rev.desc.d7.uiv.tau-T-by-type.p1.dual-d2.icon-minus-text-on-apps-with-both.n-apps\endcsname{1400.0}
\expandafter\def\csname ppt@rev.desc.d7.uiv.tau-T-by-type.p1.dual-d2.icon-minus-text-on-apps-with-both.n-apps\endcsname{$+$1400.0}
\expandafter\def\csname val@rev.desc.d7.uiv.tau-T-by-type.p1.dual-d2.icon-minus-text-on-apps-with-both.tost1.ci90-inside\endcsname{0}
\expandafter\def\csname cnt@rev.desc.d7.uiv.tau-T-by-type.p1.dual-d2.icon-minus-text-on-apps-with-both.tost1.ci90-inside\endcsname{0}
\expandafter\def\csname pct@rev.desc.d7.uiv.tau-T-by-type.p1.dual-d2.icon-minus-text-on-apps-with-both.tost1.ci90-inside\endcsname{0.0}
\expandafter\def\csname ppt@rev.desc.d7.uiv.tau-T-by-type.p1.dual-d2.icon-minus-text-on-apps-with-both.tost1.ci90-inside\endcsname{0.0}
\expandafter\def\csname val@rev.desc.d7.uiv.tau-T-by-type.p1.dual-d2.icon-minus-text-on-apps-with-both.tost1.equivalent\endcsname{0}
\expandafter\def\csname cnt@rev.desc.d7.uiv.tau-T-by-type.p1.dual-d2.icon-minus-text-on-apps-with-both.tost1.equivalent\endcsname{0}
\expandafter\def\csname pct@rev.desc.d7.uiv.tau-T-by-type.p1.dual-d2.icon-minus-text-on-apps-with-both.tost1.equivalent\endcsname{0.0}
\expandafter\def\csname ppt@rev.desc.d7.uiv.tau-T-by-type.p1.dual-d2.icon-minus-text-on-apps-with-both.tost1.equivalent\endcsname{0.0}
\expandafter\def\csname val@rev.desc.d7.uiv.tau-T-by-type.p1.dual-d2.icon-minus-text-on-apps-with-both.tost1.p\endcsname{0.597}
\expandafter\def\csname cnt@rev.desc.d7.uiv.tau-T-by-type.p1.dual-d2.icon-minus-text-on-apps-with-both.tost1.p\endcsname{1}
\expandafter\def\csname pct@rev.desc.d7.uiv.tau-T-by-type.p1.dual-d2.icon-minus-text-on-apps-with-both.tost1.p\endcsname{59.7}
\expandafter\def\csname ppt@rev.desc.d7.uiv.tau-T-by-type.p1.dual-d2.icon-minus-text-on-apps-with-both.tost1.p\endcsname{$+$59.7}
\expandafter\def\csname val@rev.desc.d7.uiv.tau-T-by-type.p1.dual-d2.icon-minus-text-on-apps-with-both.tost5.ci90-inside\endcsname{0}
\expandafter\def\csname cnt@rev.desc.d7.uiv.tau-T-by-type.p1.dual-d2.icon-minus-text-on-apps-with-both.tost5.ci90-inside\endcsname{0}
\expandafter\def\csname pct@rev.desc.d7.uiv.tau-T-by-type.p1.dual-d2.icon-minus-text-on-apps-with-both.tost5.ci90-inside\endcsname{0.0}
\expandafter\def\csname ppt@rev.desc.d7.uiv.tau-T-by-type.p1.dual-d2.icon-minus-text-on-apps-with-both.tost5.ci90-inside\endcsname{0.0}
\expandafter\def\csname val@rev.desc.d7.uiv.tau-T-by-type.p1.dual-d2.icon-minus-text-on-apps-with-both.tost5.equivalent\endcsname{0}
\expandafter\def\csname cnt@rev.desc.d7.uiv.tau-T-by-type.p1.dual-d2.icon-minus-text-on-apps-with-both.tost5.equivalent\endcsname{0}
\expandafter\def\csname pct@rev.desc.d7.uiv.tau-T-by-type.p1.dual-d2.icon-minus-text-on-apps-with-both.tost5.equivalent\endcsname{0.0}
\expandafter\def\csname ppt@rev.desc.d7.uiv.tau-T-by-type.p1.dual-d2.icon-minus-text-on-apps-with-both.tost5.equivalent\endcsname{0.0}
\expandafter\def\csname val@rev.desc.d7.uiv.tau-T-by-type.p1.dual-d2.icon-minus-text-on-apps-with-both.tost5.p\endcsname{0.356}
\expandafter\def\csname cnt@rev.desc.d7.uiv.tau-T-by-type.p1.dual-d2.icon-minus-text-on-apps-with-both.tost5.p\endcsname{0}
\expandafter\def\csname pct@rev.desc.d7.uiv.tau-T-by-type.p1.dual-d2.icon-minus-text-on-apps-with-both.tost5.p\endcsname{35.6}
\expandafter\def\csname ppt@rev.desc.d7.uiv.tau-T-by-type.p1.dual-d2.icon-minus-text-on-apps-with-both.tost5.p\endcsname{$+$35.6}
\expandafter\def\csname val@rev.desc.d7.uiv.tau-T-by-type.p1.dual-d2.icon.acc-F\endcsname{0.702}
\expandafter\def\csname cnt@rev.desc.d7.uiv.tau-T-by-type.p1.dual-d2.icon.acc-F\endcsname{1}
\expandafter\def\csname pct@rev.desc.d7.uiv.tau-T-by-type.p1.dual-d2.icon.acc-F\endcsname{70.2}
\expandafter\def\csname ppt@rev.desc.d7.uiv.tau-T-by-type.p1.dual-d2.icon.acc-F\endcsname{$+$70.2}
\expandafter\def\csname val@rev.desc.d7.uiv.tau-T-by-type.p1.dual-d2.icon.acc-direct\endcsname{0.825}
\expandafter\def\csname cnt@rev.desc.d7.uiv.tau-T-by-type.p1.dual-d2.icon.acc-direct\endcsname{1}
\expandafter\def\csname pct@rev.desc.d7.uiv.tau-T-by-type.p1.dual-d2.icon.acc-direct\endcsname{82.5}
\expandafter\def\csname ppt@rev.desc.d7.uiv.tau-T-by-type.p1.dual-d2.icon.acc-direct\endcsname{$+$82.5}
\expandafter\def\csname val@rev.desc.d7.uiv.tau-T-by-type.p1.dual-d2.icon.n-apps\endcsname{23}
\expandafter\def\csname cnt@rev.desc.d7.uiv.tau-T-by-type.p1.dual-d2.icon.n-apps\endcsname{23}
\expandafter\def\csname pct@rev.desc.d7.uiv.tau-T-by-type.p1.dual-d2.icon.n-apps\endcsname{2300.0}
\expandafter\def\csname ppt@rev.desc.d7.uiv.tau-T-by-type.p1.dual-d2.icon.n-apps\endcsname{$+$2300.0}
\expandafter\def\csname val@rev.desc.d7.uiv.tau-T-by-type.p1.dual-d2.icon.n-tasks\endcsname{114}
\expandafter\def\csname cnt@rev.desc.d7.uiv.tau-T-by-type.p1.dual-d2.icon.n-tasks\endcsname{114}
\expandafter\def\csname pct@rev.desc.d7.uiv.tau-T-by-type.p1.dual-d2.icon.n-tasks\endcsname{11400.0}
\expandafter\def\csname ppt@rev.desc.d7.uiv.tau-T-by-type.p1.dual-d2.icon.n-tasks\endcsname{$+$11400.0}
\expandafter\def\csname val@rev.desc.d7.uiv.tau-T-by-type.p1.dual-d2.icon.tau.ci90.0\endcsname{-0.162}
\expandafter\def\csname cnt@rev.desc.d7.uiv.tau-T-by-type.p1.dual-d2.icon.tau.ci90.0\endcsname{0}
\expandafter\def\csname pct@rev.desc.d7.uiv.tau-T-by-type.p1.dual-d2.icon.tau.ci90.0\endcsname{$-$16.2}
\expandafter\def\csname ppt@rev.desc.d7.uiv.tau-T-by-type.p1.dual-d2.icon.tau.ci90.0\endcsname{$-$16.2}
\expandafter\def\csname val@rev.desc.d7.uiv.tau-T-by-type.p1.dual-d2.icon.tau.ci90.1\endcsname{-0.067}
\expandafter\def\csname cnt@rev.desc.d7.uiv.tau-T-by-type.p1.dual-d2.icon.tau.ci90.1\endcsname{0}
\expandafter\def\csname pct@rev.desc.d7.uiv.tau-T-by-type.p1.dual-d2.icon.tau.ci90.1\endcsname{$-$6.7}
\expandafter\def\csname ppt@rev.desc.d7.uiv.tau-T-by-type.p1.dual-d2.icon.tau.ci90.1\endcsname{$-$6.7}
\expandafter\def\csname val@rev.desc.d7.uiv.tau-T-by-type.p1.dual-d2.icon.tau.ci95-excludes-0\endcsname{1}
\expandafter\def\csname cnt@rev.desc.d7.uiv.tau-T-by-type.p1.dual-d2.icon.tau.ci95-excludes-0\endcsname{1}
\expandafter\def\csname pct@rev.desc.d7.uiv.tau-T-by-type.p1.dual-d2.icon.tau.ci95-excludes-0\endcsname{100.0}
\expandafter\def\csname ppt@rev.desc.d7.uiv.tau-T-by-type.p1.dual-d2.icon.tau.ci95-excludes-0\endcsname{$+$100.0}
\expandafter\def\csname val@rev.desc.d7.uiv.tau-T-by-type.p1.dual-d2.icon.tau.ci95.0\endcsname{-0.171}
\expandafter\def\csname cnt@rev.desc.d7.uiv.tau-T-by-type.p1.dual-d2.icon.tau.ci95.0\endcsname{0}
\expandafter\def\csname pct@rev.desc.d7.uiv.tau-T-by-type.p1.dual-d2.icon.tau.ci95.0\endcsname{$-$17.1}
\expandafter\def\csname ppt@rev.desc.d7.uiv.tau-T-by-type.p1.dual-d2.icon.tau.ci95.0\endcsname{$-$17.1}
\expandafter\def\csname val@rev.desc.d7.uiv.tau-T-by-type.p1.dual-d2.icon.tau.ci95.1\endcsname{-0.059}
\expandafter\def\csname cnt@rev.desc.d7.uiv.tau-T-by-type.p1.dual-d2.icon.tau.ci95.1\endcsname{0}
\expandafter\def\csname pct@rev.desc.d7.uiv.tau-T-by-type.p1.dual-d2.icon.tau.ci95.1\endcsname{$-$5.9}
\expandafter\def\csname ppt@rev.desc.d7.uiv.tau-T-by-type.p1.dual-d2.icon.tau.ci95.1\endcsname{$-$5.9}
\expandafter\def\csname val@rev.desc.d7.uiv.tau-T-by-type.p1.dual-d2.icon.tau.mean\endcsname{-0.114}
\expandafter\def\csname cnt@rev.desc.d7.uiv.tau-T-by-type.p1.dual-d2.icon.tau.mean\endcsname{0}
\expandafter\def\csname pct@rev.desc.d7.uiv.tau-T-by-type.p1.dual-d2.icon.tau.mean\endcsname{$-$11.4}
\expandafter\def\csname ppt@rev.desc.d7.uiv.tau-T-by-type.p1.dual-d2.icon.tau.mean\endcsname{$-$11.4}
\expandafter\def\csname val@rev.desc.d7.uiv.tau-T-by-type.p1.dual-d2.icon.tau.n-apps\endcsname{23}
\expandafter\def\csname cnt@rev.desc.d7.uiv.tau-T-by-type.p1.dual-d2.icon.tau.n-apps\endcsname{23}
\expandafter\def\csname pct@rev.desc.d7.uiv.tau-T-by-type.p1.dual-d2.icon.tau.n-apps\endcsname{2300.0}
\expandafter\def\csname ppt@rev.desc.d7.uiv.tau-T-by-type.p1.dual-d2.icon.tau.n-apps\endcsname{$+$2300.0}
\expandafter\def\csname val@rev.desc.d7.uiv.tau-T-by-type.p1.dual-d2.icon.tau.tost1.ci90-inside\endcsname{0}
\expandafter\def\csname cnt@rev.desc.d7.uiv.tau-T-by-type.p1.dual-d2.icon.tau.tost1.ci90-inside\endcsname{0}
\expandafter\def\csname pct@rev.desc.d7.uiv.tau-T-by-type.p1.dual-d2.icon.tau.tost1.ci90-inside\endcsname{0.0}
\expandafter\def\csname ppt@rev.desc.d7.uiv.tau-T-by-type.p1.dual-d2.icon.tau.tost1.ci90-inside\endcsname{0.0}
\expandafter\def\csname val@rev.desc.d7.uiv.tau-T-by-type.p1.dual-d2.icon.tau.tost1.equivalent\endcsname{0}
\expandafter\def\csname cnt@rev.desc.d7.uiv.tau-T-by-type.p1.dual-d2.icon.tau.tost1.equivalent\endcsname{0}
\expandafter\def\csname pct@rev.desc.d7.uiv.tau-T-by-type.p1.dual-d2.icon.tau.tost1.equivalent\endcsname{0.0}
\expandafter\def\csname ppt@rev.desc.d7.uiv.tau-T-by-type.p1.dual-d2.icon.tau.tost1.equivalent\endcsname{0.0}
\expandafter\def\csname val@rev.desc.d7.uiv.tau-T-by-type.p1.dual-d2.icon.tau.tost1.p\endcsname{1.000}
\expandafter\def\csname cnt@rev.desc.d7.uiv.tau-T-by-type.p1.dual-d2.icon.tau.tost1.p\endcsname{1}
\expandafter\def\csname pct@rev.desc.d7.uiv.tau-T-by-type.p1.dual-d2.icon.tau.tost1.p\endcsname{100.0}
\expandafter\def\csname ppt@rev.desc.d7.uiv.tau-T-by-type.p1.dual-d2.icon.tau.tost1.p\endcsname{$+$100.0}
\expandafter\def\csname val@rev.desc.d7.uiv.tau-T-by-type.p1.dual-d2.icon.tau.tost5.ci90-inside\endcsname{0}
\expandafter\def\csname cnt@rev.desc.d7.uiv.tau-T-by-type.p1.dual-d2.icon.tau.tost5.ci90-inside\endcsname{0}
\expandafter\def\csname pct@rev.desc.d7.uiv.tau-T-by-type.p1.dual-d2.icon.tau.tost5.ci90-inside\endcsname{0.0}
\expandafter\def\csname ppt@rev.desc.d7.uiv.tau-T-by-type.p1.dual-d2.icon.tau.tost5.ci90-inside\endcsname{0.0}
\expandafter\def\csname val@rev.desc.d7.uiv.tau-T-by-type.p1.dual-d2.icon.tau.tost5.equivalent\endcsname{0}
\expandafter\def\csname cnt@rev.desc.d7.uiv.tau-T-by-type.p1.dual-d2.icon.tau.tost5.equivalent\endcsname{0}
\expandafter\def\csname pct@rev.desc.d7.uiv.tau-T-by-type.p1.dual-d2.icon.tau.tost5.equivalent\endcsname{0.0}
\expandafter\def\csname ppt@rev.desc.d7.uiv.tau-T-by-type.p1.dual-d2.icon.tau.tost5.equivalent\endcsname{0.0}
\expandafter\def\csname val@rev.desc.d7.uiv.tau-T-by-type.p1.dual-d2.icon.tau.tost5.p\endcsname{0.989}
\expandafter\def\csname cnt@rev.desc.d7.uiv.tau-T-by-type.p1.dual-d2.icon.tau.tost5.p\endcsname{1}
\expandafter\def\csname pct@rev.desc.d7.uiv.tau-T-by-type.p1.dual-d2.icon.tau.tost5.p\endcsname{98.9}
\expandafter\def\csname ppt@rev.desc.d7.uiv.tau-T-by-type.p1.dual-d2.icon.tau.tost5.p\endcsname{$+$98.9}
\expandafter\def\csname val@rev.desc.d7.uiv.tau-T-by-type.p1.dual-d2.text.acc-F\endcsname{0.833}
\expandafter\def\csname cnt@rev.desc.d7.uiv.tau-T-by-type.p1.dual-d2.text.acc-F\endcsname{1}
\expandafter\def\csname pct@rev.desc.d7.uiv.tau-T-by-type.p1.dual-d2.text.acc-F\endcsname{83.3}
\expandafter\def\csname ppt@rev.desc.d7.uiv.tau-T-by-type.p1.dual-d2.text.acc-F\endcsname{$+$83.3}
\expandafter\def\csname val@rev.desc.d7.uiv.tau-T-by-type.p1.dual-d2.text.acc-direct\endcsname{0.867}
\expandafter\def\csname cnt@rev.desc.d7.uiv.tau-T-by-type.p1.dual-d2.text.acc-direct\endcsname{1}
\expandafter\def\csname pct@rev.desc.d7.uiv.tau-T-by-type.p1.dual-d2.text.acc-direct\endcsname{86.7}
\expandafter\def\csname ppt@rev.desc.d7.uiv.tau-T-by-type.p1.dual-d2.text.acc-direct\endcsname{$+$86.7}
\expandafter\def\csname val@rev.desc.d7.uiv.tau-T-by-type.p1.dual-d2.text.n-apps\endcsname{15}
\expandafter\def\csname cnt@rev.desc.d7.uiv.tau-T-by-type.p1.dual-d2.text.n-apps\endcsname{15}
\expandafter\def\csname pct@rev.desc.d7.uiv.tau-T-by-type.p1.dual-d2.text.n-apps\endcsname{1500.0}
\expandafter\def\csname ppt@rev.desc.d7.uiv.tau-T-by-type.p1.dual-d2.text.n-apps\endcsname{$+$1500.0}
\expandafter\def\csname val@rev.desc.d7.uiv.tau-T-by-type.p1.dual-d2.text.n-tasks\endcsname{30}
\expandafter\def\csname cnt@rev.desc.d7.uiv.tau-T-by-type.p1.dual-d2.text.n-tasks\endcsname{30}
\expandafter\def\csname pct@rev.desc.d7.uiv.tau-T-by-type.p1.dual-d2.text.n-tasks\endcsname{3000.0}
\expandafter\def\csname ppt@rev.desc.d7.uiv.tau-T-by-type.p1.dual-d2.text.n-tasks\endcsname{$+$3000.0}
\expandafter\def\csname val@rev.desc.d7.uiv.tau-T-by-type.p1.dual-d2.text.tau.ci90.0\endcsname{-0.200}
\expandafter\def\csname cnt@rev.desc.d7.uiv.tau-T-by-type.p1.dual-d2.text.tau.ci90.0\endcsname{0}
\expandafter\def\csname pct@rev.desc.d7.uiv.tau-T-by-type.p1.dual-d2.text.tau.ci90.0\endcsname{$-$20.0}
\expandafter\def\csname ppt@rev.desc.d7.uiv.tau-T-by-type.p1.dual-d2.text.tau.ci90.0\endcsname{$-$20.0}
\expandafter\def\csname val@rev.desc.d7.uiv.tau-T-by-type.p1.dual-d2.text.tau.ci90.1\endcsname{0.000}
\expandafter\def\csname cnt@rev.desc.d7.uiv.tau-T-by-type.p1.dual-d2.text.tau.ci90.1\endcsname{0}
\expandafter\def\csname pct@rev.desc.d7.uiv.tau-T-by-type.p1.dual-d2.text.tau.ci90.1\endcsname{0.0}
\expandafter\def\csname ppt@rev.desc.d7.uiv.tau-T-by-type.p1.dual-d2.text.tau.ci90.1\endcsname{0.0}
\expandafter\def\csname val@rev.desc.d7.uiv.tau-T-by-type.p1.dual-d2.text.tau.ci95-excludes-0\endcsname{0}
\expandafter\def\csname cnt@rev.desc.d7.uiv.tau-T-by-type.p1.dual-d2.text.tau.ci95-excludes-0\endcsname{0}
\expandafter\def\csname pct@rev.desc.d7.uiv.tau-T-by-type.p1.dual-d2.text.tau.ci95-excludes-0\endcsname{0.0}
\expandafter\def\csname ppt@rev.desc.d7.uiv.tau-T-by-type.p1.dual-d2.text.tau.ci95-excludes-0\endcsname{0.0}
\expandafter\def\csname val@rev.desc.d7.uiv.tau-T-by-type.p1.dual-d2.text.tau.ci95.0\endcsname{-0.200}
\expandafter\def\csname cnt@rev.desc.d7.uiv.tau-T-by-type.p1.dual-d2.text.tau.ci95.0\endcsname{0}
\expandafter\def\csname pct@rev.desc.d7.uiv.tau-T-by-type.p1.dual-d2.text.tau.ci95.0\endcsname{$-$20.0}
\expandafter\def\csname ppt@rev.desc.d7.uiv.tau-T-by-type.p1.dual-d2.text.tau.ci95.0\endcsname{$-$20.0}
\expandafter\def\csname val@rev.desc.d7.uiv.tau-T-by-type.p1.dual-d2.text.tau.ci95.1\endcsname{0.000}
\expandafter\def\csname cnt@rev.desc.d7.uiv.tau-T-by-type.p1.dual-d2.text.tau.ci95.1\endcsname{0}
\expandafter\def\csname pct@rev.desc.d7.uiv.tau-T-by-type.p1.dual-d2.text.tau.ci95.1\endcsname{0.0}
\expandafter\def\csname ppt@rev.desc.d7.uiv.tau-T-by-type.p1.dual-d2.text.tau.ci95.1\endcsname{0.0}
\expandafter\def\csname val@rev.desc.d7.uiv.tau-T-by-type.p1.dual-d2.text.tau.mean\endcsname{-0.067}
\expandafter\def\csname cnt@rev.desc.d7.uiv.tau-T-by-type.p1.dual-d2.text.tau.mean\endcsname{0}
\expandafter\def\csname pct@rev.desc.d7.uiv.tau-T-by-type.p1.dual-d2.text.tau.mean\endcsname{$-$6.7}
\expandafter\def\csname ppt@rev.desc.d7.uiv.tau-T-by-type.p1.dual-d2.text.tau.mean\endcsname{$-$6.7}
\expandafter\def\csname val@rev.desc.d7.uiv.tau-T-by-type.p1.dual-d2.text.tau.n-apps\endcsname{15}
\expandafter\def\csname cnt@rev.desc.d7.uiv.tau-T-by-type.p1.dual-d2.text.tau.n-apps\endcsname{15}
\expandafter\def\csname pct@rev.desc.d7.uiv.tau-T-by-type.p1.dual-d2.text.tau.n-apps\endcsname{1500.0}
\expandafter\def\csname ppt@rev.desc.d7.uiv.tau-T-by-type.p1.dual-d2.text.tau.n-apps\endcsname{$+$1500.0}
\expandafter\def\csname val@rev.desc.d7.uiv.tau-T-by-type.p1.dual-d2.text.tau.tost1.ci90-inside\endcsname{0}
\expandafter\def\csname cnt@rev.desc.d7.uiv.tau-T-by-type.p1.dual-d2.text.tau.tost1.ci90-inside\endcsname{0}
\expandafter\def\csname pct@rev.desc.d7.uiv.tau-T-by-type.p1.dual-d2.text.tau.tost1.ci90-inside\endcsname{0.0}
\expandafter\def\csname ppt@rev.desc.d7.uiv.tau-T-by-type.p1.dual-d2.text.tau.tost1.ci90-inside\endcsname{0.0}
\expandafter\def\csname val@rev.desc.d7.uiv.tau-T-by-type.p1.dual-d2.text.tau.tost1.equivalent\endcsname{0}
\expandafter\def\csname cnt@rev.desc.d7.uiv.tau-T-by-type.p1.dual-d2.text.tau.tost1.equivalent\endcsname{0}
\expandafter\def\csname pct@rev.desc.d7.uiv.tau-T-by-type.p1.dual-d2.text.tau.tost1.equivalent\endcsname{0.0}
\expandafter\def\csname ppt@rev.desc.d7.uiv.tau-T-by-type.p1.dual-d2.text.tau.tost1.equivalent\endcsname{0.0}
\expandafter\def\csname val@rev.desc.d7.uiv.tau-T-by-type.p1.dual-d2.text.tau.tost1.p\endcsname{0.653}
\expandafter\def\csname cnt@rev.desc.d7.uiv.tau-T-by-type.p1.dual-d2.text.tau.tost1.p\endcsname{1}
\expandafter\def\csname pct@rev.desc.d7.uiv.tau-T-by-type.p1.dual-d2.text.tau.tost1.p\endcsname{65.3}
\expandafter\def\csname ppt@rev.desc.d7.uiv.tau-T-by-type.p1.dual-d2.text.tau.tost1.p\endcsname{$+$65.3}
\expandafter\def\csname val@rev.desc.d7.uiv.tau-T-by-type.p1.dual-d2.text.tau.tost5.ci90-inside\endcsname{0}
\expandafter\def\csname cnt@rev.desc.d7.uiv.tau-T-by-type.p1.dual-d2.text.tau.tost5.ci90-inside\endcsname{0}
\expandafter\def\csname pct@rev.desc.d7.uiv.tau-T-by-type.p1.dual-d2.text.tau.tost5.ci90-inside\endcsname{0.0}
\expandafter\def\csname ppt@rev.desc.d7.uiv.tau-T-by-type.p1.dual-d2.text.tau.tost5.ci90-inside\endcsname{0.0}
\expandafter\def\csname val@rev.desc.d7.uiv.tau-T-by-type.p1.dual-d2.text.tau.tost5.equivalent\endcsname{0}
\expandafter\def\csname cnt@rev.desc.d7.uiv.tau-T-by-type.p1.dual-d2.text.tau.tost5.equivalent\endcsname{0}
\expandafter\def\csname pct@rev.desc.d7.uiv.tau-T-by-type.p1.dual-d2.text.tau.tost5.equivalent\endcsname{0.0}
\expandafter\def\csname ppt@rev.desc.d7.uiv.tau-T-by-type.p1.dual-d2.text.tau.tost5.equivalent\endcsname{0.0}
\expandafter\def\csname val@rev.desc.d7.uiv.tau-T-by-type.p1.dual-d2.text.tau.tost5.p\endcsname{0.653}
\expandafter\def\csname cnt@rev.desc.d7.uiv.tau-T-by-type.p1.dual-d2.text.tau.tost5.p\endcsname{1}
\expandafter\def\csname pct@rev.desc.d7.uiv.tau-T-by-type.p1.dual-d2.text.tau.tost5.p\endcsname{65.3}
\expandafter\def\csname ppt@rev.desc.d7.uiv.tau-T-by-type.p1.dual-d2.text.tau.tost5.p\endcsname{$+$65.3}
\expandafter\def\csname val@rev.desc.d7.uiv.tau-T-by-type.p1.grid2-d3.icon-minus-text-on-apps-with-both.ci90.0\endcsname{-0.058}
\expandafter\def\csname cnt@rev.desc.d7.uiv.tau-T-by-type.p1.grid2-d3.icon-minus-text-on-apps-with-both.ci90.0\endcsname{0}
\expandafter\def\csname pct@rev.desc.d7.uiv.tau-T-by-type.p1.grid2-d3.icon-minus-text-on-apps-with-both.ci90.0\endcsname{$-$5.8}
\expandafter\def\csname ppt@rev.desc.d7.uiv.tau-T-by-type.p1.grid2-d3.icon-minus-text-on-apps-with-both.ci90.0\endcsname{$-$5.8}
\expandafter\def\csname val@rev.desc.d7.uiv.tau-T-by-type.p1.grid2-d3.icon-minus-text-on-apps-with-both.ci90.1\endcsname{0.188}
\expandafter\def\csname cnt@rev.desc.d7.uiv.tau-T-by-type.p1.grid2-d3.icon-minus-text-on-apps-with-both.ci90.1\endcsname{0}
\expandafter\def\csname pct@rev.desc.d7.uiv.tau-T-by-type.p1.grid2-d3.icon-minus-text-on-apps-with-both.ci90.1\endcsname{18.8}
\expandafter\def\csname ppt@rev.desc.d7.uiv.tau-T-by-type.p1.grid2-d3.icon-minus-text-on-apps-with-both.ci90.1\endcsname{$+$18.8}
\expandafter\def\csname val@rev.desc.d7.uiv.tau-T-by-type.p1.grid2-d3.icon-minus-text-on-apps-with-both.ci95-excludes-0\endcsname{0}
\expandafter\def\csname cnt@rev.desc.d7.uiv.tau-T-by-type.p1.grid2-d3.icon-minus-text-on-apps-with-both.ci95-excludes-0\endcsname{0}
\expandafter\def\csname pct@rev.desc.d7.uiv.tau-T-by-type.p1.grid2-d3.icon-minus-text-on-apps-with-both.ci95-excludes-0\endcsname{0.0}
\expandafter\def\csname ppt@rev.desc.d7.uiv.tau-T-by-type.p1.grid2-d3.icon-minus-text-on-apps-with-both.ci95-excludes-0\endcsname{0.0}
\expandafter\def\csname val@rev.desc.d7.uiv.tau-T-by-type.p1.grid2-d3.icon-minus-text-on-apps-with-both.ci95.0\endcsname{-0.082}
\expandafter\def\csname cnt@rev.desc.d7.uiv.tau-T-by-type.p1.grid2-d3.icon-minus-text-on-apps-with-both.ci95.0\endcsname{0}
\expandafter\def\csname pct@rev.desc.d7.uiv.tau-T-by-type.p1.grid2-d3.icon-minus-text-on-apps-with-both.ci95.0\endcsname{$-$8.2}
\expandafter\def\csname ppt@rev.desc.d7.uiv.tau-T-by-type.p1.grid2-d3.icon-minus-text-on-apps-with-both.ci95.0\endcsname{$-$8.2}
\expandafter\def\csname val@rev.desc.d7.uiv.tau-T-by-type.p1.grid2-d3.icon-minus-text-on-apps-with-both.ci95.1\endcsname{0.210}
\expandafter\def\csname cnt@rev.desc.d7.uiv.tau-T-by-type.p1.grid2-d3.icon-minus-text-on-apps-with-both.ci95.1\endcsname{0}
\expandafter\def\csname pct@rev.desc.d7.uiv.tau-T-by-type.p1.grid2-d3.icon-minus-text-on-apps-with-both.ci95.1\endcsname{21.0}
\expandafter\def\csname ppt@rev.desc.d7.uiv.tau-T-by-type.p1.grid2-d3.icon-minus-text-on-apps-with-both.ci95.1\endcsname{$+$21.0}
\expandafter\def\csname val@rev.desc.d7.uiv.tau-T-by-type.p1.grid2-d3.icon-minus-text-on-apps-with-both.mean\endcsname{0.056}
\expandafter\def\csname cnt@rev.desc.d7.uiv.tau-T-by-type.p1.grid2-d3.icon-minus-text-on-apps-with-both.mean\endcsname{0}
\expandafter\def\csname pct@rev.desc.d7.uiv.tau-T-by-type.p1.grid2-d3.icon-minus-text-on-apps-with-both.mean\endcsname{5.6}
\expandafter\def\csname ppt@rev.desc.d7.uiv.tau-T-by-type.p1.grid2-d3.icon-minus-text-on-apps-with-both.mean\endcsname{$+$5.6}
\expandafter\def\csname val@rev.desc.d7.uiv.tau-T-by-type.p1.grid2-d3.icon-minus-text-on-apps-with-both.n-apps\endcsname{14}
\expandafter\def\csname cnt@rev.desc.d7.uiv.tau-T-by-type.p1.grid2-d3.icon-minus-text-on-apps-with-both.n-apps\endcsname{14}
\expandafter\def\csname pct@rev.desc.d7.uiv.tau-T-by-type.p1.grid2-d3.icon-minus-text-on-apps-with-both.n-apps\endcsname{1400.0}
\expandafter\def\csname ppt@rev.desc.d7.uiv.tau-T-by-type.p1.grid2-d3.icon-minus-text-on-apps-with-both.n-apps\endcsname{$+$1400.0}
\expandafter\def\csname val@rev.desc.d7.uiv.tau-T-by-type.p1.grid2-d3.icon-minus-text-on-apps-with-both.tost1.ci90-inside\endcsname{0}
\expandafter\def\csname cnt@rev.desc.d7.uiv.tau-T-by-type.p1.grid2-d3.icon-minus-text-on-apps-with-both.tost1.ci90-inside\endcsname{0}
\expandafter\def\csname pct@rev.desc.d7.uiv.tau-T-by-type.p1.grid2-d3.icon-minus-text-on-apps-with-both.tost1.ci90-inside\endcsname{0.0}
\expandafter\def\csname ppt@rev.desc.d7.uiv.tau-T-by-type.p1.grid2-d3.icon-minus-text-on-apps-with-both.tost1.ci90-inside\endcsname{0.0}
\expandafter\def\csname val@rev.desc.d7.uiv.tau-T-by-type.p1.grid2-d3.icon-minus-text-on-apps-with-both.tost1.equivalent\endcsname{0}
\expandafter\def\csname cnt@rev.desc.d7.uiv.tau-T-by-type.p1.grid2-d3.icon-minus-text-on-apps-with-both.tost1.equivalent\endcsname{0}
\expandafter\def\csname pct@rev.desc.d7.uiv.tau-T-by-type.p1.grid2-d3.icon-minus-text-on-apps-with-both.tost1.equivalent\endcsname{0.0}
\expandafter\def\csname ppt@rev.desc.d7.uiv.tau-T-by-type.p1.grid2-d3.icon-minus-text-on-apps-with-both.tost1.equivalent\endcsname{0.0}
\expandafter\def\csname val@rev.desc.d7.uiv.tau-T-by-type.p1.grid2-d3.icon-minus-text-on-apps-with-both.tost1.p\endcsname{0.715}
\expandafter\def\csname cnt@rev.desc.d7.uiv.tau-T-by-type.p1.grid2-d3.icon-minus-text-on-apps-with-both.tost1.p\endcsname{1}
\expandafter\def\csname pct@rev.desc.d7.uiv.tau-T-by-type.p1.grid2-d3.icon-minus-text-on-apps-with-both.tost1.p\endcsname{71.5}
\expandafter\def\csname ppt@rev.desc.d7.uiv.tau-T-by-type.p1.grid2-d3.icon-minus-text-on-apps-with-both.tost1.p\endcsname{$+$71.5}
\expandafter\def\csname val@rev.desc.d7.uiv.tau-T-by-type.p1.grid2-d3.icon-minus-text-on-apps-with-both.tost5.ci90-inside\endcsname{0}
\expandafter\def\csname cnt@rev.desc.d7.uiv.tau-T-by-type.p1.grid2-d3.icon-minus-text-on-apps-with-both.tost5.ci90-inside\endcsname{0}
\expandafter\def\csname pct@rev.desc.d7.uiv.tau-T-by-type.p1.grid2-d3.icon-minus-text-on-apps-with-both.tost5.ci90-inside\endcsname{0.0}
\expandafter\def\csname ppt@rev.desc.d7.uiv.tau-T-by-type.p1.grid2-d3.icon-minus-text-on-apps-with-both.tost5.ci90-inside\endcsname{0.0}
\expandafter\def\csname val@rev.desc.d7.uiv.tau-T-by-type.p1.grid2-d3.icon-minus-text-on-apps-with-both.tost5.equivalent\endcsname{0}
\expandafter\def\csname cnt@rev.desc.d7.uiv.tau-T-by-type.p1.grid2-d3.icon-minus-text-on-apps-with-both.tost5.equivalent\endcsname{0}
\expandafter\def\csname pct@rev.desc.d7.uiv.tau-T-by-type.p1.grid2-d3.icon-minus-text-on-apps-with-both.tost5.equivalent\endcsname{0.0}
\expandafter\def\csname ppt@rev.desc.d7.uiv.tau-T-by-type.p1.grid2-d3.icon-minus-text-on-apps-with-both.tost5.equivalent\endcsname{0.0}
\expandafter\def\csname val@rev.desc.d7.uiv.tau-T-by-type.p1.grid2-d3.icon-minus-text-on-apps-with-both.tost5.p\endcsname{0.514}
\expandafter\def\csname cnt@rev.desc.d7.uiv.tau-T-by-type.p1.grid2-d3.icon-minus-text-on-apps-with-both.tost5.p\endcsname{1}
\expandafter\def\csname pct@rev.desc.d7.uiv.tau-T-by-type.p1.grid2-d3.icon-minus-text-on-apps-with-both.tost5.p\endcsname{51.4}
\expandafter\def\csname ppt@rev.desc.d7.uiv.tau-T-by-type.p1.grid2-d3.icon-minus-text-on-apps-with-both.tost5.p\endcsname{$+$51.4}
\expandafter\def\csname val@rev.desc.d7.uiv.tau-T-by-type.p1.grid2-d3.icon.acc-F\endcsname{0.746}
\expandafter\def\csname cnt@rev.desc.d7.uiv.tau-T-by-type.p1.grid2-d3.icon.acc-F\endcsname{1}
\expandafter\def\csname pct@rev.desc.d7.uiv.tau-T-by-type.p1.grid2-d3.icon.acc-F\endcsname{74.6}
\expandafter\def\csname ppt@rev.desc.d7.uiv.tau-T-by-type.p1.grid2-d3.icon.acc-F\endcsname{$+$74.6}
\expandafter\def\csname val@rev.desc.d7.uiv.tau-T-by-type.p1.grid2-d3.icon.acc-direct\endcsname{0.825}
\expandafter\def\csname cnt@rev.desc.d7.uiv.tau-T-by-type.p1.grid2-d3.icon.acc-direct\endcsname{1}
\expandafter\def\csname pct@rev.desc.d7.uiv.tau-T-by-type.p1.grid2-d3.icon.acc-direct\endcsname{82.5}
\expandafter\def\csname ppt@rev.desc.d7.uiv.tau-T-by-type.p1.grid2-d3.icon.acc-direct\endcsname{$+$82.5}
\expandafter\def\csname val@rev.desc.d7.uiv.tau-T-by-type.p1.grid2-d3.icon.n-apps\endcsname{23}
\expandafter\def\csname cnt@rev.desc.d7.uiv.tau-T-by-type.p1.grid2-d3.icon.n-apps\endcsname{23}
\expandafter\def\csname pct@rev.desc.d7.uiv.tau-T-by-type.p1.grid2-d3.icon.n-apps\endcsname{2300.0}
\expandafter\def\csname ppt@rev.desc.d7.uiv.tau-T-by-type.p1.grid2-d3.icon.n-apps\endcsname{$+$2300.0}
\expandafter\def\csname val@rev.desc.d7.uiv.tau-T-by-type.p1.grid2-d3.icon.n-tasks\endcsname{114}
\expandafter\def\csname cnt@rev.desc.d7.uiv.tau-T-by-type.p1.grid2-d3.icon.n-tasks\endcsname{114}
\expandafter\def\csname pct@rev.desc.d7.uiv.tau-T-by-type.p1.grid2-d3.icon.n-tasks\endcsname{11400.0}
\expandafter\def\csname ppt@rev.desc.d7.uiv.tau-T-by-type.p1.grid2-d3.icon.n-tasks\endcsname{$+$11400.0}
\expandafter\def\csname val@rev.desc.d7.uiv.tau-T-by-type.p1.grid2-d3.icon.tau.ci90.0\endcsname{-0.122}
\expandafter\def\csname cnt@rev.desc.d7.uiv.tau-T-by-type.p1.grid2-d3.icon.tau.ci90.0\endcsname{0}
\expandafter\def\csname pct@rev.desc.d7.uiv.tau-T-by-type.p1.grid2-d3.icon.tau.ci90.0\endcsname{$-$12.2}
\expandafter\def\csname ppt@rev.desc.d7.uiv.tau-T-by-type.p1.grid2-d3.icon.tau.ci90.0\endcsname{$-$12.2}
\expandafter\def\csname val@rev.desc.d7.uiv.tau-T-by-type.p1.grid2-d3.icon.tau.ci90.1\endcsname{-0.004}
\expandafter\def\csname cnt@rev.desc.d7.uiv.tau-T-by-type.p1.grid2-d3.icon.tau.ci90.1\endcsname{0}
\expandafter\def\csname pct@rev.desc.d7.uiv.tau-T-by-type.p1.grid2-d3.icon.tau.ci90.1\endcsname{$-$0.4}
\expandafter\def\csname ppt@rev.desc.d7.uiv.tau-T-by-type.p1.grid2-d3.icon.tau.ci90.1\endcsname{$-$0.4}
\expandafter\def\csname val@rev.desc.d7.uiv.tau-T-by-type.p1.grid2-d3.icon.tau.ci95-excludes-0\endcsname{0}
\expandafter\def\csname cnt@rev.desc.d7.uiv.tau-T-by-type.p1.grid2-d3.icon.tau.ci95-excludes-0\endcsname{0}
\expandafter\def\csname pct@rev.desc.d7.uiv.tau-T-by-type.p1.grid2-d3.icon.tau.ci95-excludes-0\endcsname{0.0}
\expandafter\def\csname ppt@rev.desc.d7.uiv.tau-T-by-type.p1.grid2-d3.icon.tau.ci95-excludes-0\endcsname{0.0}
\expandafter\def\csname val@rev.desc.d7.uiv.tau-T-by-type.p1.grid2-d3.icon.tau.ci95.0\endcsname{-0.132}
\expandafter\def\csname cnt@rev.desc.d7.uiv.tau-T-by-type.p1.grid2-d3.icon.tau.ci95.0\endcsname{0}
\expandafter\def\csname pct@rev.desc.d7.uiv.tau-T-by-type.p1.grid2-d3.icon.tau.ci95.0\endcsname{$-$13.2}
\expandafter\def\csname ppt@rev.desc.d7.uiv.tau-T-by-type.p1.grid2-d3.icon.tau.ci95.0\endcsname{$-$13.2}
\expandafter\def\csname val@rev.desc.d7.uiv.tau-T-by-type.p1.grid2-d3.icon.tau.ci95.1\endcsname{0.007}
\expandafter\def\csname cnt@rev.desc.d7.uiv.tau-T-by-type.p1.grid2-d3.icon.tau.ci95.1\endcsname{0}
\expandafter\def\csname pct@rev.desc.d7.uiv.tau-T-by-type.p1.grid2-d3.icon.tau.ci95.1\endcsname{0.7}
\expandafter\def\csname ppt@rev.desc.d7.uiv.tau-T-by-type.p1.grid2-d3.icon.tau.ci95.1\endcsname{$+$0.7}
\expandafter\def\csname val@rev.desc.d7.uiv.tau-T-by-type.p1.grid2-d3.icon.tau.mean\endcsname{-0.064}
\expandafter\def\csname cnt@rev.desc.d7.uiv.tau-T-by-type.p1.grid2-d3.icon.tau.mean\endcsname{0}
\expandafter\def\csname pct@rev.desc.d7.uiv.tau-T-by-type.p1.grid2-d3.icon.tau.mean\endcsname{$-$6.4}
\expandafter\def\csname ppt@rev.desc.d7.uiv.tau-T-by-type.p1.grid2-d3.icon.tau.mean\endcsname{$-$6.4}
\expandafter\def\csname val@rev.desc.d7.uiv.tau-T-by-type.p1.grid2-d3.icon.tau.n-apps\endcsname{23}
\expandafter\def\csname cnt@rev.desc.d7.uiv.tau-T-by-type.p1.grid2-d3.icon.tau.n-apps\endcsname{23}
\expandafter\def\csname pct@rev.desc.d7.uiv.tau-T-by-type.p1.grid2-d3.icon.tau.n-apps\endcsname{2300.0}
\expandafter\def\csname ppt@rev.desc.d7.uiv.tau-T-by-type.p1.grid2-d3.icon.tau.n-apps\endcsname{$+$2300.0}
\expandafter\def\csname val@rev.desc.d7.uiv.tau-T-by-type.p1.grid2-d3.icon.tau.tost1.ci90-inside\endcsname{0}
\expandafter\def\csname cnt@rev.desc.d7.uiv.tau-T-by-type.p1.grid2-d3.icon.tau.tost1.ci90-inside\endcsname{0}
\expandafter\def\csname pct@rev.desc.d7.uiv.tau-T-by-type.p1.grid2-d3.icon.tau.tost1.ci90-inside\endcsname{0.0}
\expandafter\def\csname ppt@rev.desc.d7.uiv.tau-T-by-type.p1.grid2-d3.icon.tau.tost1.ci90-inside\endcsname{0.0}
\expandafter\def\csname val@rev.desc.d7.uiv.tau-T-by-type.p1.grid2-d3.icon.tau.tost1.equivalent\endcsname{0}
\expandafter\def\csname cnt@rev.desc.d7.uiv.tau-T-by-type.p1.grid2-d3.icon.tau.tost1.equivalent\endcsname{0}
\expandafter\def\csname pct@rev.desc.d7.uiv.tau-T-by-type.p1.grid2-d3.icon.tau.tost1.equivalent\endcsname{0.0}
\expandafter\def\csname ppt@rev.desc.d7.uiv.tau-T-by-type.p1.grid2-d3.icon.tau.tost1.equivalent\endcsname{0.0}
\expandafter\def\csname val@rev.desc.d7.uiv.tau-T-by-type.p1.grid2-d3.icon.tau.tost1.p\endcsname{0.933}
\expandafter\def\csname cnt@rev.desc.d7.uiv.tau-T-by-type.p1.grid2-d3.icon.tau.tost1.p\endcsname{1}
\expandafter\def\csname pct@rev.desc.d7.uiv.tau-T-by-type.p1.grid2-d3.icon.tau.tost1.p\endcsname{93.3}
\expandafter\def\csname ppt@rev.desc.d7.uiv.tau-T-by-type.p1.grid2-d3.icon.tau.tost1.p\endcsname{$+$93.3}
\expandafter\def\csname val@rev.desc.d7.uiv.tau-T-by-type.p1.grid2-d3.icon.tau.tost5.ci90-inside\endcsname{0}
\expandafter\def\csname cnt@rev.desc.d7.uiv.tau-T-by-type.p1.grid2-d3.icon.tau.tost5.ci90-inside\endcsname{0}
\expandafter\def\csname pct@rev.desc.d7.uiv.tau-T-by-type.p1.grid2-d3.icon.tau.tost5.ci90-inside\endcsname{0.0}
\expandafter\def\csname ppt@rev.desc.d7.uiv.tau-T-by-type.p1.grid2-d3.icon.tau.tost5.ci90-inside\endcsname{0.0}
\expandafter\def\csname val@rev.desc.d7.uiv.tau-T-by-type.p1.grid2-d3.icon.tau.tost5.equivalent\endcsname{0}
\expandafter\def\csname cnt@rev.desc.d7.uiv.tau-T-by-type.p1.grid2-d3.icon.tau.tost5.equivalent\endcsname{0}
\expandafter\def\csname pct@rev.desc.d7.uiv.tau-T-by-type.p1.grid2-d3.icon.tau.tost5.equivalent\endcsname{0.0}
\expandafter\def\csname ppt@rev.desc.d7.uiv.tau-T-by-type.p1.grid2-d3.icon.tau.tost5.equivalent\endcsname{0.0}
\expandafter\def\csname val@rev.desc.d7.uiv.tau-T-by-type.p1.grid2-d3.icon.tau.tost5.p\endcsname{0.652}
\expandafter\def\csname cnt@rev.desc.d7.uiv.tau-T-by-type.p1.grid2-d3.icon.tau.tost5.p\endcsname{1}
\expandafter\def\csname pct@rev.desc.d7.uiv.tau-T-by-type.p1.grid2-d3.icon.tau.tost5.p\endcsname{65.2}
\expandafter\def\csname ppt@rev.desc.d7.uiv.tau-T-by-type.p1.grid2-d3.icon.tau.tost5.p\endcsname{$+$65.2}
\expandafter\def\csname val@rev.desc.d7.uiv.tau-T-by-type.p1.grid2-d3.text.acc-F\endcsname{0.800}
\expandafter\def\csname cnt@rev.desc.d7.uiv.tau-T-by-type.p1.grid2-d3.text.acc-F\endcsname{1}
\expandafter\def\csname pct@rev.desc.d7.uiv.tau-T-by-type.p1.grid2-d3.text.acc-F\endcsname{80.0}
\expandafter\def\csname ppt@rev.desc.d7.uiv.tau-T-by-type.p1.grid2-d3.text.acc-F\endcsname{$+$80.0}
\expandafter\def\csname val@rev.desc.d7.uiv.tau-T-by-type.p1.grid2-d3.text.acc-direct\endcsname{0.867}
\expandafter\def\csname cnt@rev.desc.d7.uiv.tau-T-by-type.p1.grid2-d3.text.acc-direct\endcsname{1}
\expandafter\def\csname pct@rev.desc.d7.uiv.tau-T-by-type.p1.grid2-d3.text.acc-direct\endcsname{86.7}
\expandafter\def\csname ppt@rev.desc.d7.uiv.tau-T-by-type.p1.grid2-d3.text.acc-direct\endcsname{$+$86.7}
\expandafter\def\csname val@rev.desc.d7.uiv.tau-T-by-type.p1.grid2-d3.text.n-apps\endcsname{15}
\expandafter\def\csname cnt@rev.desc.d7.uiv.tau-T-by-type.p1.grid2-d3.text.n-apps\endcsname{15}
\expandafter\def\csname pct@rev.desc.d7.uiv.tau-T-by-type.p1.grid2-d3.text.n-apps\endcsname{1500.0}
\expandafter\def\csname ppt@rev.desc.d7.uiv.tau-T-by-type.p1.grid2-d3.text.n-apps\endcsname{$+$1500.0}
\expandafter\def\csname val@rev.desc.d7.uiv.tau-T-by-type.p1.grid2-d3.text.n-tasks\endcsname{30}
\expandafter\def\csname cnt@rev.desc.d7.uiv.tau-T-by-type.p1.grid2-d3.text.n-tasks\endcsname{30}
\expandafter\def\csname pct@rev.desc.d7.uiv.tau-T-by-type.p1.grid2-d3.text.n-tasks\endcsname{3000.0}
\expandafter\def\csname ppt@rev.desc.d7.uiv.tau-T-by-type.p1.grid2-d3.text.n-tasks\endcsname{$+$3000.0}
\expandafter\def\csname val@rev.desc.d7.uiv.tau-T-by-type.p1.grid2-d3.text.tau.ci90.0\endcsname{-0.200}
\expandafter\def\csname cnt@rev.desc.d7.uiv.tau-T-by-type.p1.grid2-d3.text.tau.ci90.0\endcsname{0}
\expandafter\def\csname pct@rev.desc.d7.uiv.tau-T-by-type.p1.grid2-d3.text.tau.ci90.0\endcsname{$-$20.0}
\expandafter\def\csname ppt@rev.desc.d7.uiv.tau-T-by-type.p1.grid2-d3.text.tau.ci90.0\endcsname{$-$20.0}
\expandafter\def\csname val@rev.desc.d7.uiv.tau-T-by-type.p1.grid2-d3.text.tau.ci90.1\endcsname{0.000}
\expandafter\def\csname cnt@rev.desc.d7.uiv.tau-T-by-type.p1.grid2-d3.text.tau.ci90.1\endcsname{0}
\expandafter\def\csname pct@rev.desc.d7.uiv.tau-T-by-type.p1.grid2-d3.text.tau.ci90.1\endcsname{0.0}
\expandafter\def\csname ppt@rev.desc.d7.uiv.tau-T-by-type.p1.grid2-d3.text.tau.ci90.1\endcsname{0.0}
\expandafter\def\csname val@rev.desc.d7.uiv.tau-T-by-type.p1.grid2-d3.text.tau.ci95-excludes-0\endcsname{0}
\expandafter\def\csname cnt@rev.desc.d7.uiv.tau-T-by-type.p1.grid2-d3.text.tau.ci95-excludes-0\endcsname{0}
\expandafter\def\csname pct@rev.desc.d7.uiv.tau-T-by-type.p1.grid2-d3.text.tau.ci95-excludes-0\endcsname{0.0}
\expandafter\def\csname ppt@rev.desc.d7.uiv.tau-T-by-type.p1.grid2-d3.text.tau.ci95-excludes-0\endcsname{0.0}
\expandafter\def\csname val@rev.desc.d7.uiv.tau-T-by-type.p1.grid2-d3.text.tau.ci95.0\endcsname{-0.233}
\expandafter\def\csname cnt@rev.desc.d7.uiv.tau-T-by-type.p1.grid2-d3.text.tau.ci95.0\endcsname{0}
\expandafter\def\csname pct@rev.desc.d7.uiv.tau-T-by-type.p1.grid2-d3.text.tau.ci95.0\endcsname{$-$23.3}
\expandafter\def\csname ppt@rev.desc.d7.uiv.tau-T-by-type.p1.grid2-d3.text.tau.ci95.0\endcsname{$-$23.3}
\expandafter\def\csname val@rev.desc.d7.uiv.tau-T-by-type.p1.grid2-d3.text.tau.ci95.1\endcsname{0.000}
\expandafter\def\csname cnt@rev.desc.d7.uiv.tau-T-by-type.p1.grid2-d3.text.tau.ci95.1\endcsname{0}
\expandafter\def\csname pct@rev.desc.d7.uiv.tau-T-by-type.p1.grid2-d3.text.tau.ci95.1\endcsname{0.0}
\expandafter\def\csname ppt@rev.desc.d7.uiv.tau-T-by-type.p1.grid2-d3.text.tau.ci95.1\endcsname{0.0}
\expandafter\def\csname val@rev.desc.d7.uiv.tau-T-by-type.p1.grid2-d3.text.tau.mean\endcsname{-0.083}
\expandafter\def\csname cnt@rev.desc.d7.uiv.tau-T-by-type.p1.grid2-d3.text.tau.mean\endcsname{0}
\expandafter\def\csname pct@rev.desc.d7.uiv.tau-T-by-type.p1.grid2-d3.text.tau.mean\endcsname{$-$8.3}
\expandafter\def\csname ppt@rev.desc.d7.uiv.tau-T-by-type.p1.grid2-d3.text.tau.mean\endcsname{$-$8.3}
\expandafter\def\csname val@rev.desc.d7.uiv.tau-T-by-type.p1.grid2-d3.text.tau.n-apps\endcsname{15}
\expandafter\def\csname cnt@rev.desc.d7.uiv.tau-T-by-type.p1.grid2-d3.text.tau.n-apps\endcsname{15}
\expandafter\def\csname pct@rev.desc.d7.uiv.tau-T-by-type.p1.grid2-d3.text.tau.n-apps\endcsname{1500.0}
\expandafter\def\csname ppt@rev.desc.d7.uiv.tau-T-by-type.p1.grid2-d3.text.tau.n-apps\endcsname{$+$1500.0}
\expandafter\def\csname val@rev.desc.d7.uiv.tau-T-by-type.p1.grid2-d3.text.tau.tost1.ci90-inside\endcsname{0}
\expandafter\def\csname cnt@rev.desc.d7.uiv.tau-T-by-type.p1.grid2-d3.text.tau.tost1.ci90-inside\endcsname{0}
\expandafter\def\csname pct@rev.desc.d7.uiv.tau-T-by-type.p1.grid2-d3.text.tau.tost1.ci90-inside\endcsname{0.0}
\expandafter\def\csname ppt@rev.desc.d7.uiv.tau-T-by-type.p1.grid2-d3.text.tau.tost1.ci90-inside\endcsname{0.0}
\expandafter\def\csname val@rev.desc.d7.uiv.tau-T-by-type.p1.grid2-d3.text.tau.tost1.equivalent\endcsname{0}
\expandafter\def\csname cnt@rev.desc.d7.uiv.tau-T-by-type.p1.grid2-d3.text.tau.tost1.equivalent\endcsname{0}
\expandafter\def\csname pct@rev.desc.d7.uiv.tau-T-by-type.p1.grid2-d3.text.tau.tost1.equivalent\endcsname{0.0}
\expandafter\def\csname ppt@rev.desc.d7.uiv.tau-T-by-type.p1.grid2-d3.text.tau.tost1.equivalent\endcsname{0.0}
\expandafter\def\csname val@rev.desc.d7.uiv.tau-T-by-type.p1.grid2-d3.text.tau.tost1.p\endcsname{0.888}
\expandafter\def\csname cnt@rev.desc.d7.uiv.tau-T-by-type.p1.grid2-d3.text.tau.tost1.p\endcsname{1}
\expandafter\def\csname pct@rev.desc.d7.uiv.tau-T-by-type.p1.grid2-d3.text.tau.tost1.p\endcsname{88.8}
\expandafter\def\csname ppt@rev.desc.d7.uiv.tau-T-by-type.p1.grid2-d3.text.tau.tost1.p\endcsname{$+$88.8}
\expandafter\def\csname val@rev.desc.d7.uiv.tau-T-by-type.p1.grid2-d3.text.tau.tost5.ci90-inside\endcsname{0}
\expandafter\def\csname cnt@rev.desc.d7.uiv.tau-T-by-type.p1.grid2-d3.text.tau.tost5.ci90-inside\endcsname{0}
\expandafter\def\csname pct@rev.desc.d7.uiv.tau-T-by-type.p1.grid2-d3.text.tau.tost5.ci90-inside\endcsname{0.0}
\expandafter\def\csname ppt@rev.desc.d7.uiv.tau-T-by-type.p1.grid2-d3.text.tau.tost5.ci90-inside\endcsname{0.0}
\expandafter\def\csname val@rev.desc.d7.uiv.tau-T-by-type.p1.grid2-d3.text.tau.tost5.equivalent\endcsname{0}
\expandafter\def\csname cnt@rev.desc.d7.uiv.tau-T-by-type.p1.grid2-d3.text.tau.tost5.equivalent\endcsname{0}
\expandafter\def\csname pct@rev.desc.d7.uiv.tau-T-by-type.p1.grid2-d3.text.tau.tost5.equivalent\endcsname{0.0}
\expandafter\def\csname ppt@rev.desc.d7.uiv.tau-T-by-type.p1.grid2-d3.text.tau.tost5.equivalent\endcsname{0.0}
\expandafter\def\csname val@rev.desc.d7.uiv.tau-T-by-type.p1.grid2-d3.text.tau.tost5.p\endcsname{0.680}
\expandafter\def\csname cnt@rev.desc.d7.uiv.tau-T-by-type.p1.grid2-d3.text.tau.tost5.p\endcsname{1}
\expandafter\def\csname pct@rev.desc.d7.uiv.tau-T-by-type.p1.grid2-d3.text.tau.tost5.p\endcsname{68.0}
\expandafter\def\csname ppt@rev.desc.d7.uiv.tau-T-by-type.p1.grid2-d3.text.tau.tost5.p\endcsname{$+$68.0}
\expandafter\def\csname val@rev.desc.d7.uiv.tau-T-by-type.p1.zoomclick-d3.icon-minus-text-on-apps-with-both.ci90.0\endcsname{-0.079}
\expandafter\def\csname cnt@rev.desc.d7.uiv.tau-T-by-type.p1.zoomclick-d3.icon-minus-text-on-apps-with-both.ci90.0\endcsname{0}
\expandafter\def\csname pct@rev.desc.d7.uiv.tau-T-by-type.p1.zoomclick-d3.icon-minus-text-on-apps-with-both.ci90.0\endcsname{$-$7.9}
\expandafter\def\csname ppt@rev.desc.d7.uiv.tau-T-by-type.p1.zoomclick-d3.icon-minus-text-on-apps-with-both.ci90.0\endcsname{$-$7.9}
\expandafter\def\csname val@rev.desc.d7.uiv.tau-T-by-type.p1.zoomclick-d3.icon-minus-text-on-apps-with-both.ci90.1\endcsname{0.129}
\expandafter\def\csname cnt@rev.desc.d7.uiv.tau-T-by-type.p1.zoomclick-d3.icon-minus-text-on-apps-with-both.ci90.1\endcsname{0}
\expandafter\def\csname pct@rev.desc.d7.uiv.tau-T-by-type.p1.zoomclick-d3.icon-minus-text-on-apps-with-both.ci90.1\endcsname{12.9}
\expandafter\def\csname ppt@rev.desc.d7.uiv.tau-T-by-type.p1.zoomclick-d3.icon-minus-text-on-apps-with-both.ci90.1\endcsname{$+$12.9}
\expandafter\def\csname val@rev.desc.d7.uiv.tau-T-by-type.p1.zoomclick-d3.icon-minus-text-on-apps-with-both.ci95-excludes-0\endcsname{0}
\expandafter\def\csname cnt@rev.desc.d7.uiv.tau-T-by-type.p1.zoomclick-d3.icon-minus-text-on-apps-with-both.ci95-excludes-0\endcsname{0}
\expandafter\def\csname pct@rev.desc.d7.uiv.tau-T-by-type.p1.zoomclick-d3.icon-minus-text-on-apps-with-both.ci95-excludes-0\endcsname{0.0}
\expandafter\def\csname ppt@rev.desc.d7.uiv.tau-T-by-type.p1.zoomclick-d3.icon-minus-text-on-apps-with-both.ci95-excludes-0\endcsname{0.0}
\expandafter\def\csname val@rev.desc.d7.uiv.tau-T-by-type.p1.zoomclick-d3.icon-minus-text-on-apps-with-both.ci95.0\endcsname{-0.089}
\expandafter\def\csname cnt@rev.desc.d7.uiv.tau-T-by-type.p1.zoomclick-d3.icon-minus-text-on-apps-with-both.ci95.0\endcsname{0}
\expandafter\def\csname pct@rev.desc.d7.uiv.tau-T-by-type.p1.zoomclick-d3.icon-minus-text-on-apps-with-both.ci95.0\endcsname{$-$8.9}
\expandafter\def\csname ppt@rev.desc.d7.uiv.tau-T-by-type.p1.zoomclick-d3.icon-minus-text-on-apps-with-both.ci95.0\endcsname{$-$8.9}
\expandafter\def\csname val@rev.desc.d7.uiv.tau-T-by-type.p1.zoomclick-d3.icon-minus-text-on-apps-with-both.ci95.1\endcsname{0.157}
\expandafter\def\csname cnt@rev.desc.d7.uiv.tau-T-by-type.p1.zoomclick-d3.icon-minus-text-on-apps-with-both.ci95.1\endcsname{0}
\expandafter\def\csname pct@rev.desc.d7.uiv.tau-T-by-type.p1.zoomclick-d3.icon-minus-text-on-apps-with-both.ci95.1\endcsname{15.7}
\expandafter\def\csname ppt@rev.desc.d7.uiv.tau-T-by-type.p1.zoomclick-d3.icon-minus-text-on-apps-with-both.ci95.1\endcsname{$+$15.7}
\expandafter\def\csname val@rev.desc.d7.uiv.tau-T-by-type.p1.zoomclick-d3.icon-minus-text-on-apps-with-both.mean\endcsname{0.011}
\expandafter\def\csname cnt@rev.desc.d7.uiv.tau-T-by-type.p1.zoomclick-d3.icon-minus-text-on-apps-with-both.mean\endcsname{0}
\expandafter\def\csname pct@rev.desc.d7.uiv.tau-T-by-type.p1.zoomclick-d3.icon-minus-text-on-apps-with-both.mean\endcsname{1.1}
\expandafter\def\csname ppt@rev.desc.d7.uiv.tau-T-by-type.p1.zoomclick-d3.icon-minus-text-on-apps-with-both.mean\endcsname{$+$1.1}
\expandafter\def\csname val@rev.desc.d7.uiv.tau-T-by-type.p1.zoomclick-d3.icon-minus-text-on-apps-with-both.n-apps\endcsname{14}
\expandafter\def\csname cnt@rev.desc.d7.uiv.tau-T-by-type.p1.zoomclick-d3.icon-minus-text-on-apps-with-both.n-apps\endcsname{14}
\expandafter\def\csname pct@rev.desc.d7.uiv.tau-T-by-type.p1.zoomclick-d3.icon-minus-text-on-apps-with-both.n-apps\endcsname{1400.0}
\expandafter\def\csname ppt@rev.desc.d7.uiv.tau-T-by-type.p1.zoomclick-d3.icon-minus-text-on-apps-with-both.n-apps\endcsname{$+$1400.0}
\expandafter\def\csname val@rev.desc.d7.uiv.tau-T-by-type.p1.zoomclick-d3.icon-minus-text-on-apps-with-both.tost1.ci90-inside\endcsname{0}
\expandafter\def\csname cnt@rev.desc.d7.uiv.tau-T-by-type.p1.zoomclick-d3.icon-minus-text-on-apps-with-both.tost1.ci90-inside\endcsname{0}
\expandafter\def\csname pct@rev.desc.d7.uiv.tau-T-by-type.p1.zoomclick-d3.icon-minus-text-on-apps-with-both.tost1.ci90-inside\endcsname{0.0}
\expandafter\def\csname ppt@rev.desc.d7.uiv.tau-T-by-type.p1.zoomclick-d3.icon-minus-text-on-apps-with-both.tost1.ci90-inside\endcsname{0.0}
\expandafter\def\csname val@rev.desc.d7.uiv.tau-T-by-type.p1.zoomclick-d3.icon-minus-text-on-apps-with-both.tost1.equivalent\endcsname{0}
\expandafter\def\csname cnt@rev.desc.d7.uiv.tau-T-by-type.p1.zoomclick-d3.icon-minus-text-on-apps-with-both.tost1.equivalent\endcsname{0}
\expandafter\def\csname pct@rev.desc.d7.uiv.tau-T-by-type.p1.zoomclick-d3.icon-minus-text-on-apps-with-both.tost1.equivalent\endcsname{0.0}
\expandafter\def\csname ppt@rev.desc.d7.uiv.tau-T-by-type.p1.zoomclick-d3.icon-minus-text-on-apps-with-both.tost1.equivalent\endcsname{0.0}
\expandafter\def\csname val@rev.desc.d7.uiv.tau-T-by-type.p1.zoomclick-d3.icon-minus-text-on-apps-with-both.tost1.p\endcsname{0.505}
\expandafter\def\csname cnt@rev.desc.d7.uiv.tau-T-by-type.p1.zoomclick-d3.icon-minus-text-on-apps-with-both.tost1.p\endcsname{1}
\expandafter\def\csname pct@rev.desc.d7.uiv.tau-T-by-type.p1.zoomclick-d3.icon-minus-text-on-apps-with-both.tost1.p\endcsname{50.5}
\expandafter\def\csname ppt@rev.desc.d7.uiv.tau-T-by-type.p1.zoomclick-d3.icon-minus-text-on-apps-with-both.tost1.p\endcsname{$+$50.5}
\expandafter\def\csname val@rev.desc.d7.uiv.tau-T-by-type.p1.zoomclick-d3.icon-minus-text-on-apps-with-both.tost5.ci90-inside\endcsname{0}
\expandafter\def\csname cnt@rev.desc.d7.uiv.tau-T-by-type.p1.zoomclick-d3.icon-minus-text-on-apps-with-both.tost5.ci90-inside\endcsname{0}
\expandafter\def\csname pct@rev.desc.d7.uiv.tau-T-by-type.p1.zoomclick-d3.icon-minus-text-on-apps-with-both.tost5.ci90-inside\endcsname{0.0}
\expandafter\def\csname ppt@rev.desc.d7.uiv.tau-T-by-type.p1.zoomclick-d3.icon-minus-text-on-apps-with-both.tost5.ci90-inside\endcsname{0.0}
\expandafter\def\csname val@rev.desc.d7.uiv.tau-T-by-type.p1.zoomclick-d3.icon-minus-text-on-apps-with-both.tost5.equivalent\endcsname{0}
\expandafter\def\csname cnt@rev.desc.d7.uiv.tau-T-by-type.p1.zoomclick-d3.icon-minus-text-on-apps-with-both.tost5.equivalent\endcsname{0}
\expandafter\def\csname pct@rev.desc.d7.uiv.tau-T-by-type.p1.zoomclick-d3.icon-minus-text-on-apps-with-both.tost5.equivalent\endcsname{0.0}
\expandafter\def\csname ppt@rev.desc.d7.uiv.tau-T-by-type.p1.zoomclick-d3.icon-minus-text-on-apps-with-both.tost5.equivalent\endcsname{0.0}
\expandafter\def\csname val@rev.desc.d7.uiv.tau-T-by-type.p1.zoomclick-d3.icon-minus-text-on-apps-with-both.tost5.p\endcsname{0.259}
\expandafter\def\csname cnt@rev.desc.d7.uiv.tau-T-by-type.p1.zoomclick-d3.icon-minus-text-on-apps-with-both.tost5.p\endcsname{0}
\expandafter\def\csname pct@rev.desc.d7.uiv.tau-T-by-type.p1.zoomclick-d3.icon-minus-text-on-apps-with-both.tost5.p\endcsname{25.9}
\expandafter\def\csname ppt@rev.desc.d7.uiv.tau-T-by-type.p1.zoomclick-d3.icon-minus-text-on-apps-with-both.tost5.p\endcsname{$+$25.9}
\expandafter\def\csname val@rev.desc.d7.uiv.tau-T-by-type.p1.zoomclick-d3.icon.acc-F\endcsname{0.763}
\expandafter\def\csname cnt@rev.desc.d7.uiv.tau-T-by-type.p1.zoomclick-d3.icon.acc-F\endcsname{1}
\expandafter\def\csname pct@rev.desc.d7.uiv.tau-T-by-type.p1.zoomclick-d3.icon.acc-F\endcsname{76.3}
\expandafter\def\csname ppt@rev.desc.d7.uiv.tau-T-by-type.p1.zoomclick-d3.icon.acc-F\endcsname{$+$76.3}
\expandafter\def\csname val@rev.desc.d7.uiv.tau-T-by-type.p1.zoomclick-d3.icon.acc-direct\endcsname{0.825}
\expandafter\def\csname cnt@rev.desc.d7.uiv.tau-T-by-type.p1.zoomclick-d3.icon.acc-direct\endcsname{1}
\expandafter\def\csname pct@rev.desc.d7.uiv.tau-T-by-type.p1.zoomclick-d3.icon.acc-direct\endcsname{82.5}
\expandafter\def\csname ppt@rev.desc.d7.uiv.tau-T-by-type.p1.zoomclick-d3.icon.acc-direct\endcsname{$+$82.5}
\expandafter\def\csname val@rev.desc.d7.uiv.tau-T-by-type.p1.zoomclick-d3.icon.n-apps\endcsname{23}
\expandafter\def\csname cnt@rev.desc.d7.uiv.tau-T-by-type.p1.zoomclick-d3.icon.n-apps\endcsname{23}
\expandafter\def\csname pct@rev.desc.d7.uiv.tau-T-by-type.p1.zoomclick-d3.icon.n-apps\endcsname{2300.0}
\expandafter\def\csname ppt@rev.desc.d7.uiv.tau-T-by-type.p1.zoomclick-d3.icon.n-apps\endcsname{$+$2300.0}
\expandafter\def\csname val@rev.desc.d7.uiv.tau-T-by-type.p1.zoomclick-d3.icon.n-tasks\endcsname{114}
\expandafter\def\csname cnt@rev.desc.d7.uiv.tau-T-by-type.p1.zoomclick-d3.icon.n-tasks\endcsname{114}
\expandafter\def\csname pct@rev.desc.d7.uiv.tau-T-by-type.p1.zoomclick-d3.icon.n-tasks\endcsname{11400.0}
\expandafter\def\csname ppt@rev.desc.d7.uiv.tau-T-by-type.p1.zoomclick-d3.icon.n-tasks\endcsname{$+$11400.0}
\expandafter\def\csname val@rev.desc.d7.uiv.tau-T-by-type.p1.zoomclick-d3.icon.tau.ci90.0\endcsname{-0.109}
\expandafter\def\csname cnt@rev.desc.d7.uiv.tau-T-by-type.p1.zoomclick-d3.icon.tau.ci90.0\endcsname{0}
\expandafter\def\csname pct@rev.desc.d7.uiv.tau-T-by-type.p1.zoomclick-d3.icon.tau.ci90.0\endcsname{$-$10.9}
\expandafter\def\csname ppt@rev.desc.d7.uiv.tau-T-by-type.p1.zoomclick-d3.icon.tau.ci90.0\endcsname{$-$10.9}
\expandafter\def\csname val@rev.desc.d7.uiv.tau-T-by-type.p1.zoomclick-d3.icon.tau.ci90.1\endcsname{-0.017}
\expandafter\def\csname cnt@rev.desc.d7.uiv.tau-T-by-type.p1.zoomclick-d3.icon.tau.ci90.1\endcsname{0}
\expandafter\def\csname pct@rev.desc.d7.uiv.tau-T-by-type.p1.zoomclick-d3.icon.tau.ci90.1\endcsname{$-$1.7}
\expandafter\def\csname ppt@rev.desc.d7.uiv.tau-T-by-type.p1.zoomclick-d3.icon.tau.ci90.1\endcsname{$-$1.7}
\expandafter\def\csname val@rev.desc.d7.uiv.tau-T-by-type.p1.zoomclick-d3.icon.tau.ci95-excludes-0\endcsname{1}
\expandafter\def\csname cnt@rev.desc.d7.uiv.tau-T-by-type.p1.zoomclick-d3.icon.tau.ci95-excludes-0\endcsname{1}
\expandafter\def\csname pct@rev.desc.d7.uiv.tau-T-by-type.p1.zoomclick-d3.icon.tau.ci95-excludes-0\endcsname{100.0}
\expandafter\def\csname ppt@rev.desc.d7.uiv.tau-T-by-type.p1.zoomclick-d3.icon.tau.ci95-excludes-0\endcsname{$+$100.0}
\expandafter\def\csname val@rev.desc.d7.uiv.tau-T-by-type.p1.zoomclick-d3.icon.tau.ci95.0\endcsname{-0.120}
\expandafter\def\csname cnt@rev.desc.d7.uiv.tau-T-by-type.p1.zoomclick-d3.icon.tau.ci95.0\endcsname{0}
\expandafter\def\csname pct@rev.desc.d7.uiv.tau-T-by-type.p1.zoomclick-d3.icon.tau.ci95.0\endcsname{$-$12.0}
\expandafter\def\csname ppt@rev.desc.d7.uiv.tau-T-by-type.p1.zoomclick-d3.icon.tau.ci95.0\endcsname{$-$12.0}
\expandafter\def\csname val@rev.desc.d7.uiv.tau-T-by-type.p1.zoomclick-d3.icon.tau.ci95.1\endcsname{-0.010}
\expandafter\def\csname cnt@rev.desc.d7.uiv.tau-T-by-type.p1.zoomclick-d3.icon.tau.ci95.1\endcsname{0}
\expandafter\def\csname pct@rev.desc.d7.uiv.tau-T-by-type.p1.zoomclick-d3.icon.tau.ci95.1\endcsname{$-$1.0}
\expandafter\def\csname ppt@rev.desc.d7.uiv.tau-T-by-type.p1.zoomclick-d3.icon.tau.ci95.1\endcsname{$-$1.0}
\expandafter\def\csname val@rev.desc.d7.uiv.tau-T-by-type.p1.zoomclick-d3.icon.tau.mean\endcsname{-0.059}
\expandafter\def\csname cnt@rev.desc.d7.uiv.tau-T-by-type.p1.zoomclick-d3.icon.tau.mean\endcsname{0}
\expandafter\def\csname pct@rev.desc.d7.uiv.tau-T-by-type.p1.zoomclick-d3.icon.tau.mean\endcsname{$-$5.9}
\expandafter\def\csname ppt@rev.desc.d7.uiv.tau-T-by-type.p1.zoomclick-d3.icon.tau.mean\endcsname{$-$5.9}
\expandafter\def\csname val@rev.desc.d7.uiv.tau-T-by-type.p1.zoomclick-d3.icon.tau.n-apps\endcsname{23}
\expandafter\def\csname cnt@rev.desc.d7.uiv.tau-T-by-type.p1.zoomclick-d3.icon.tau.n-apps\endcsname{23}
\expandafter\def\csname pct@rev.desc.d7.uiv.tau-T-by-type.p1.zoomclick-d3.icon.tau.n-apps\endcsname{2300.0}
\expandafter\def\csname ppt@rev.desc.d7.uiv.tau-T-by-type.p1.zoomclick-d3.icon.tau.n-apps\endcsname{$+$2300.0}
\expandafter\def\csname val@rev.desc.d7.uiv.tau-T-by-type.p1.zoomclick-d3.icon.tau.tost1.ci90-inside\endcsname{0}
\expandafter\def\csname cnt@rev.desc.d7.uiv.tau-T-by-type.p1.zoomclick-d3.icon.tau.tost1.ci90-inside\endcsname{0}
\expandafter\def\csname pct@rev.desc.d7.uiv.tau-T-by-type.p1.zoomclick-d3.icon.tau.tost1.ci90-inside\endcsname{0.0}
\expandafter\def\csname ppt@rev.desc.d7.uiv.tau-T-by-type.p1.zoomclick-d3.icon.tau.tost1.ci90-inside\endcsname{0.0}
\expandafter\def\csname val@rev.desc.d7.uiv.tau-T-by-type.p1.zoomclick-d3.icon.tau.tost1.equivalent\endcsname{0}
\expandafter\def\csname cnt@rev.desc.d7.uiv.tau-T-by-type.p1.zoomclick-d3.icon.tau.tost1.equivalent\endcsname{0}
\expandafter\def\csname pct@rev.desc.d7.uiv.tau-T-by-type.p1.zoomclick-d3.icon.tau.tost1.equivalent\endcsname{0.0}
\expandafter\def\csname ppt@rev.desc.d7.uiv.tau-T-by-type.p1.zoomclick-d3.icon.tau.tost1.equivalent\endcsname{0.0}
\expandafter\def\csname val@rev.desc.d7.uiv.tau-T-by-type.p1.zoomclick-d3.icon.tau.tost1.p\endcsname{0.978}
\expandafter\def\csname cnt@rev.desc.d7.uiv.tau-T-by-type.p1.zoomclick-d3.icon.tau.tost1.p\endcsname{1}
\expandafter\def\csname pct@rev.desc.d7.uiv.tau-T-by-type.p1.zoomclick-d3.icon.tau.tost1.p\endcsname{97.8}
\expandafter\def\csname ppt@rev.desc.d7.uiv.tau-T-by-type.p1.zoomclick-d3.icon.tau.tost1.p\endcsname{$+$97.8}
\expandafter\def\csname val@rev.desc.d7.uiv.tau-T-by-type.p1.zoomclick-d3.icon.tau.tost5.ci90-inside\endcsname{0}
\expandafter\def\csname cnt@rev.desc.d7.uiv.tau-T-by-type.p1.zoomclick-d3.icon.tau.tost5.ci90-inside\endcsname{0}
\expandafter\def\csname pct@rev.desc.d7.uiv.tau-T-by-type.p1.zoomclick-d3.icon.tau.tost5.ci90-inside\endcsname{0.0}
\expandafter\def\csname ppt@rev.desc.d7.uiv.tau-T-by-type.p1.zoomclick-d3.icon.tau.tost5.ci90-inside\endcsname{0.0}
\expandafter\def\csname val@rev.desc.d7.uiv.tau-T-by-type.p1.zoomclick-d3.icon.tau.tost5.equivalent\endcsname{0}
\expandafter\def\csname cnt@rev.desc.d7.uiv.tau-T-by-type.p1.zoomclick-d3.icon.tau.tost5.equivalent\endcsname{0}
\expandafter\def\csname pct@rev.desc.d7.uiv.tau-T-by-type.p1.zoomclick-d3.icon.tau.tost5.equivalent\endcsname{0.0}
\expandafter\def\csname ppt@rev.desc.d7.uiv.tau-T-by-type.p1.zoomclick-d3.icon.tau.tost5.equivalent\endcsname{0.0}
\expandafter\def\csname val@rev.desc.d7.uiv.tau-T-by-type.p1.zoomclick-d3.icon.tau.tost5.p\endcsname{0.607}
\expandafter\def\csname cnt@rev.desc.d7.uiv.tau-T-by-type.p1.zoomclick-d3.icon.tau.tost5.p\endcsname{1}
\expandafter\def\csname pct@rev.desc.d7.uiv.tau-T-by-type.p1.zoomclick-d3.icon.tau.tost5.p\endcsname{60.7}
\expandafter\def\csname ppt@rev.desc.d7.uiv.tau-T-by-type.p1.zoomclick-d3.icon.tau.tost5.p\endcsname{$+$60.7}
\expandafter\def\csname val@rev.desc.d7.uiv.tau-T-by-type.p1.zoomclick-d3.text.acc-F\endcsname{0.833}
\expandafter\def\csname cnt@rev.desc.d7.uiv.tau-T-by-type.p1.zoomclick-d3.text.acc-F\endcsname{1}
\expandafter\def\csname pct@rev.desc.d7.uiv.tau-T-by-type.p1.zoomclick-d3.text.acc-F\endcsname{83.3}
\expandafter\def\csname ppt@rev.desc.d7.uiv.tau-T-by-type.p1.zoomclick-d3.text.acc-F\endcsname{$+$83.3}
\expandafter\def\csname val@rev.desc.d7.uiv.tau-T-by-type.p1.zoomclick-d3.text.acc-direct\endcsname{0.867}
\expandafter\def\csname cnt@rev.desc.d7.uiv.tau-T-by-type.p1.zoomclick-d3.text.acc-direct\endcsname{1}
\expandafter\def\csname pct@rev.desc.d7.uiv.tau-T-by-type.p1.zoomclick-d3.text.acc-direct\endcsname{86.7}
\expandafter\def\csname ppt@rev.desc.d7.uiv.tau-T-by-type.p1.zoomclick-d3.text.acc-direct\endcsname{$+$86.7}
\expandafter\def\csname val@rev.desc.d7.uiv.tau-T-by-type.p1.zoomclick-d3.text.n-apps\endcsname{15}
\expandafter\def\csname cnt@rev.desc.d7.uiv.tau-T-by-type.p1.zoomclick-d3.text.n-apps\endcsname{15}
\expandafter\def\csname pct@rev.desc.d7.uiv.tau-T-by-type.p1.zoomclick-d3.text.n-apps\endcsname{1500.0}
\expandafter\def\csname ppt@rev.desc.d7.uiv.tau-T-by-type.p1.zoomclick-d3.text.n-apps\endcsname{$+$1500.0}
\expandafter\def\csname val@rev.desc.d7.uiv.tau-T-by-type.p1.zoomclick-d3.text.n-tasks\endcsname{30}
\expandafter\def\csname cnt@rev.desc.d7.uiv.tau-T-by-type.p1.zoomclick-d3.text.n-tasks\endcsname{30}
\expandafter\def\csname pct@rev.desc.d7.uiv.tau-T-by-type.p1.zoomclick-d3.text.n-tasks\endcsname{3000.0}
\expandafter\def\csname ppt@rev.desc.d7.uiv.tau-T-by-type.p1.zoomclick-d3.text.n-tasks\endcsname{$+$3000.0}
\expandafter\def\csname val@rev.desc.d7.uiv.tau-T-by-type.p1.zoomclick-d3.text.tau.ci90.0\endcsname{-0.200}
\expandafter\def\csname cnt@rev.desc.d7.uiv.tau-T-by-type.p1.zoomclick-d3.text.tau.ci90.0\endcsname{0}
\expandafter\def\csname pct@rev.desc.d7.uiv.tau-T-by-type.p1.zoomclick-d3.text.tau.ci90.0\endcsname{$-$20.0}
\expandafter\def\csname ppt@rev.desc.d7.uiv.tau-T-by-type.p1.zoomclick-d3.text.tau.ci90.0\endcsname{$-$20.0}
\expandafter\def\csname val@rev.desc.d7.uiv.tau-T-by-type.p1.zoomclick-d3.text.tau.ci90.1\endcsname{0.000}
\expandafter\def\csname cnt@rev.desc.d7.uiv.tau-T-by-type.p1.zoomclick-d3.text.tau.ci90.1\endcsname{0}
\expandafter\def\csname pct@rev.desc.d7.uiv.tau-T-by-type.p1.zoomclick-d3.text.tau.ci90.1\endcsname{0.0}
\expandafter\def\csname ppt@rev.desc.d7.uiv.tau-T-by-type.p1.zoomclick-d3.text.tau.ci90.1\endcsname{0.0}
\expandafter\def\csname val@rev.desc.d7.uiv.tau-T-by-type.p1.zoomclick-d3.text.tau.ci95-excludes-0\endcsname{0}
\expandafter\def\csname cnt@rev.desc.d7.uiv.tau-T-by-type.p1.zoomclick-d3.text.tau.ci95-excludes-0\endcsname{0}
\expandafter\def\csname pct@rev.desc.d7.uiv.tau-T-by-type.p1.zoomclick-d3.text.tau.ci95-excludes-0\endcsname{0.0}
\expandafter\def\csname ppt@rev.desc.d7.uiv.tau-T-by-type.p1.zoomclick-d3.text.tau.ci95-excludes-0\endcsname{0.0}
\expandafter\def\csname val@rev.desc.d7.uiv.tau-T-by-type.p1.zoomclick-d3.text.tau.ci95.0\endcsname{-0.200}
\expandafter\def\csname cnt@rev.desc.d7.uiv.tau-T-by-type.p1.zoomclick-d3.text.tau.ci95.0\endcsname{0}
\expandafter\def\csname pct@rev.desc.d7.uiv.tau-T-by-type.p1.zoomclick-d3.text.tau.ci95.0\endcsname{$-$20.0}
\expandafter\def\csname ppt@rev.desc.d7.uiv.tau-T-by-type.p1.zoomclick-d3.text.tau.ci95.0\endcsname{$-$20.0}
\expandafter\def\csname val@rev.desc.d7.uiv.tau-T-by-type.p1.zoomclick-d3.text.tau.ci95.1\endcsname{0.000}
\expandafter\def\csname cnt@rev.desc.d7.uiv.tau-T-by-type.p1.zoomclick-d3.text.tau.ci95.1\endcsname{0}
\expandafter\def\csname pct@rev.desc.d7.uiv.tau-T-by-type.p1.zoomclick-d3.text.tau.ci95.1\endcsname{0.0}
\expandafter\def\csname ppt@rev.desc.d7.uiv.tau-T-by-type.p1.zoomclick-d3.text.tau.ci95.1\endcsname{0.0}
\expandafter\def\csname val@rev.desc.d7.uiv.tau-T-by-type.p1.zoomclick-d3.text.tau.mean\endcsname{-0.067}
\expandafter\def\csname cnt@rev.desc.d7.uiv.tau-T-by-type.p1.zoomclick-d3.text.tau.mean\endcsname{0}
\expandafter\def\csname pct@rev.desc.d7.uiv.tau-T-by-type.p1.zoomclick-d3.text.tau.mean\endcsname{$-$6.7}
\expandafter\def\csname ppt@rev.desc.d7.uiv.tau-T-by-type.p1.zoomclick-d3.text.tau.mean\endcsname{$-$6.7}
\expandafter\def\csname val@rev.desc.d7.uiv.tau-T-by-type.p1.zoomclick-d3.text.tau.n-apps\endcsname{15}
\expandafter\def\csname cnt@rev.desc.d7.uiv.tau-T-by-type.p1.zoomclick-d3.text.tau.n-apps\endcsname{15}
\expandafter\def\csname pct@rev.desc.d7.uiv.tau-T-by-type.p1.zoomclick-d3.text.tau.n-apps\endcsname{1500.0}
\expandafter\def\csname ppt@rev.desc.d7.uiv.tau-T-by-type.p1.zoomclick-d3.text.tau.n-apps\endcsname{$+$1500.0}
\expandafter\def\csname val@rev.desc.d7.uiv.tau-T-by-type.p1.zoomclick-d3.text.tau.tost1.ci90-inside\endcsname{0}
\expandafter\def\csname cnt@rev.desc.d7.uiv.tau-T-by-type.p1.zoomclick-d3.text.tau.tost1.ci90-inside\endcsname{0}
\expandafter\def\csname pct@rev.desc.d7.uiv.tau-T-by-type.p1.zoomclick-d3.text.tau.tost1.ci90-inside\endcsname{0.0}
\expandafter\def\csname ppt@rev.desc.d7.uiv.tau-T-by-type.p1.zoomclick-d3.text.tau.tost1.ci90-inside\endcsname{0.0}
\expandafter\def\csname val@rev.desc.d7.uiv.tau-T-by-type.p1.zoomclick-d3.text.tau.tost1.equivalent\endcsname{0}
\expandafter\def\csname cnt@rev.desc.d7.uiv.tau-T-by-type.p1.zoomclick-d3.text.tau.tost1.equivalent\endcsname{0}
\expandafter\def\csname pct@rev.desc.d7.uiv.tau-T-by-type.p1.zoomclick-d3.text.tau.tost1.equivalent\endcsname{0.0}
\expandafter\def\csname ppt@rev.desc.d7.uiv.tau-T-by-type.p1.zoomclick-d3.text.tau.tost1.equivalent\endcsname{0.0}
\expandafter\def\csname val@rev.desc.d7.uiv.tau-T-by-type.p1.zoomclick-d3.text.tau.tost1.p\endcsname{0.653}
\expandafter\def\csname cnt@rev.desc.d7.uiv.tau-T-by-type.p1.zoomclick-d3.text.tau.tost1.p\endcsname{1}
\expandafter\def\csname pct@rev.desc.d7.uiv.tau-T-by-type.p1.zoomclick-d3.text.tau.tost1.p\endcsname{65.3}
\expandafter\def\csname ppt@rev.desc.d7.uiv.tau-T-by-type.p1.zoomclick-d3.text.tau.tost1.p\endcsname{$+$65.3}
\expandafter\def\csname val@rev.desc.d7.uiv.tau-T-by-type.p1.zoomclick-d3.text.tau.tost5.ci90-inside\endcsname{0}
\expandafter\def\csname cnt@rev.desc.d7.uiv.tau-T-by-type.p1.zoomclick-d3.text.tau.tost5.ci90-inside\endcsname{0}
\expandafter\def\csname pct@rev.desc.d7.uiv.tau-T-by-type.p1.zoomclick-d3.text.tau.tost5.ci90-inside\endcsname{0.0}
\expandafter\def\csname ppt@rev.desc.d7.uiv.tau-T-by-type.p1.zoomclick-d3.text.tau.tost5.ci90-inside\endcsname{0.0}
\expandafter\def\csname val@rev.desc.d7.uiv.tau-T-by-type.p1.zoomclick-d3.text.tau.tost5.equivalent\endcsname{0}
\expandafter\def\csname cnt@rev.desc.d7.uiv.tau-T-by-type.p1.zoomclick-d3.text.tau.tost5.equivalent\endcsname{0}
\expandafter\def\csname pct@rev.desc.d7.uiv.tau-T-by-type.p1.zoomclick-d3.text.tau.tost5.equivalent\endcsname{0.0}
\expandafter\def\csname ppt@rev.desc.d7.uiv.tau-T-by-type.p1.zoomclick-d3.text.tau.tost5.equivalent\endcsname{0.0}
\expandafter\def\csname val@rev.desc.d7.uiv.tau-T-by-type.p1.zoomclick-d3.text.tau.tost5.p\endcsname{0.653}
\expandafter\def\csname cnt@rev.desc.d7.uiv.tau-T-by-type.p1.zoomclick-d3.text.tau.tost5.p\endcsname{1}
\expandafter\def\csname pct@rev.desc.d7.uiv.tau-T-by-type.p1.zoomclick-d3.text.tau.tost5.p\endcsname{65.3}
\expandafter\def\csname ppt@rev.desc.d7.uiv.tau-T-by-type.p1.zoomclick-d3.text.tau.tost5.p\endcsname{$+$65.3}
\expandafter\def\csname val@rev.desc.d7.uiv.tau-T-by-type.p2.center-d1.icon-minus-text-on-apps-with-both.ci90.0\endcsname{0.000}
\expandafter\def\csname cnt@rev.desc.d7.uiv.tau-T-by-type.p2.center-d1.icon-minus-text-on-apps-with-both.ci90.0\endcsname{0}
\expandafter\def\csname pct@rev.desc.d7.uiv.tau-T-by-type.p2.center-d1.icon-minus-text-on-apps-with-both.ci90.0\endcsname{0.0}
\expandafter\def\csname ppt@rev.desc.d7.uiv.tau-T-by-type.p2.center-d1.icon-minus-text-on-apps-with-both.ci90.0\endcsname{0.0}
\expandafter\def\csname val@rev.desc.d7.uiv.tau-T-by-type.p2.center-d1.icon-minus-text-on-apps-with-both.ci90.1\endcsname{0.043}
\expandafter\def\csname cnt@rev.desc.d7.uiv.tau-T-by-type.p2.center-d1.icon-minus-text-on-apps-with-both.ci90.1\endcsname{0}
\expandafter\def\csname pct@rev.desc.d7.uiv.tau-T-by-type.p2.center-d1.icon-minus-text-on-apps-with-both.ci90.1\endcsname{4.3}
\expandafter\def\csname ppt@rev.desc.d7.uiv.tau-T-by-type.p2.center-d1.icon-minus-text-on-apps-with-both.ci90.1\endcsname{$+$4.3}
\expandafter\def\csname val@rev.desc.d7.uiv.tau-T-by-type.p2.center-d1.icon-minus-text-on-apps-with-both.ci95-excludes-0\endcsname{0}
\expandafter\def\csname cnt@rev.desc.d7.uiv.tau-T-by-type.p2.center-d1.icon-minus-text-on-apps-with-both.ci95-excludes-0\endcsname{0}
\expandafter\def\csname pct@rev.desc.d7.uiv.tau-T-by-type.p2.center-d1.icon-minus-text-on-apps-with-both.ci95-excludes-0\endcsname{0.0}
\expandafter\def\csname ppt@rev.desc.d7.uiv.tau-T-by-type.p2.center-d1.icon-minus-text-on-apps-with-both.ci95-excludes-0\endcsname{0.0}
\expandafter\def\csname val@rev.desc.d7.uiv.tau-T-by-type.p2.center-d1.icon-minus-text-on-apps-with-both.ci95.0\endcsname{0.000}
\expandafter\def\csname cnt@rev.desc.d7.uiv.tau-T-by-type.p2.center-d1.icon-minus-text-on-apps-with-both.ci95.0\endcsname{0}
\expandafter\def\csname pct@rev.desc.d7.uiv.tau-T-by-type.p2.center-d1.icon-minus-text-on-apps-with-both.ci95.0\endcsname{0.0}
\expandafter\def\csname ppt@rev.desc.d7.uiv.tau-T-by-type.p2.center-d1.icon-minus-text-on-apps-with-both.ci95.0\endcsname{0.0}
\expandafter\def\csname val@rev.desc.d7.uiv.tau-T-by-type.p2.center-d1.icon-minus-text-on-apps-with-both.ci95.1\endcsname{0.043}
\expandafter\def\csname cnt@rev.desc.d7.uiv.tau-T-by-type.p2.center-d1.icon-minus-text-on-apps-with-both.ci95.1\endcsname{0}
\expandafter\def\csname pct@rev.desc.d7.uiv.tau-T-by-type.p2.center-d1.icon-minus-text-on-apps-with-both.ci95.1\endcsname{4.3}
\expandafter\def\csname ppt@rev.desc.d7.uiv.tau-T-by-type.p2.center-d1.icon-minus-text-on-apps-with-both.ci95.1\endcsname{$+$4.3}
\expandafter\def\csname val@rev.desc.d7.uiv.tau-T-by-type.p2.center-d1.icon-minus-text-on-apps-with-both.mean\endcsname{0.014}
\expandafter\def\csname cnt@rev.desc.d7.uiv.tau-T-by-type.p2.center-d1.icon-minus-text-on-apps-with-both.mean\endcsname{0}
\expandafter\def\csname pct@rev.desc.d7.uiv.tau-T-by-type.p2.center-d1.icon-minus-text-on-apps-with-both.mean\endcsname{1.4}
\expandafter\def\csname ppt@rev.desc.d7.uiv.tau-T-by-type.p2.center-d1.icon-minus-text-on-apps-with-both.mean\endcsname{$+$1.4}
\expandafter\def\csname val@rev.desc.d7.uiv.tau-T-by-type.p2.center-d1.icon-minus-text-on-apps-with-both.n-apps\endcsname{14}
\expandafter\def\csname cnt@rev.desc.d7.uiv.tau-T-by-type.p2.center-d1.icon-minus-text-on-apps-with-both.n-apps\endcsname{14}
\expandafter\def\csname pct@rev.desc.d7.uiv.tau-T-by-type.p2.center-d1.icon-minus-text-on-apps-with-both.n-apps\endcsname{1400.0}
\expandafter\def\csname ppt@rev.desc.d7.uiv.tau-T-by-type.p2.center-d1.icon-minus-text-on-apps-with-both.n-apps\endcsname{$+$1400.0}
\expandafter\def\csname val@rev.desc.d7.uiv.tau-T-by-type.p2.center-d1.icon-minus-text-on-apps-with-both.tost1.ci90-inside\endcsname{0}
\expandafter\def\csname cnt@rev.desc.d7.uiv.tau-T-by-type.p2.center-d1.icon-minus-text-on-apps-with-both.tost1.ci90-inside\endcsname{0}
\expandafter\def\csname pct@rev.desc.d7.uiv.tau-T-by-type.p2.center-d1.icon-minus-text-on-apps-with-both.tost1.ci90-inside\endcsname{0.0}
\expandafter\def\csname ppt@rev.desc.d7.uiv.tau-T-by-type.p2.center-d1.icon-minus-text-on-apps-with-both.tost1.ci90-inside\endcsname{0.0}
\expandafter\def\csname val@rev.desc.d7.uiv.tau-T-by-type.p2.center-d1.icon-minus-text-on-apps-with-both.tost1.equivalent\endcsname{0}
\expandafter\def\csname cnt@rev.desc.d7.uiv.tau-T-by-type.p2.center-d1.icon-minus-text-on-apps-with-both.tost1.equivalent\endcsname{0}
\expandafter\def\csname pct@rev.desc.d7.uiv.tau-T-by-type.p2.center-d1.icon-minus-text-on-apps-with-both.tost1.equivalent\endcsname{0.0}
\expandafter\def\csname ppt@rev.desc.d7.uiv.tau-T-by-type.p2.center-d1.icon-minus-text-on-apps-with-both.tost1.equivalent\endcsname{0.0}
\expandafter\def\csname val@rev.desc.d7.uiv.tau-T-by-type.p2.center-d1.icon-minus-text-on-apps-with-both.tost1.p\endcsname{0.653}
\expandafter\def\csname cnt@rev.desc.d7.uiv.tau-T-by-type.p2.center-d1.icon-minus-text-on-apps-with-both.tost1.p\endcsname{1}
\expandafter\def\csname pct@rev.desc.d7.uiv.tau-T-by-type.p2.center-d1.icon-minus-text-on-apps-with-both.tost1.p\endcsname{65.3}
\expandafter\def\csname ppt@rev.desc.d7.uiv.tau-T-by-type.p2.center-d1.icon-minus-text-on-apps-with-both.tost1.p\endcsname{$+$65.3}
\expandafter\def\csname val@rev.desc.d7.uiv.tau-T-by-type.p2.center-d1.icon-minus-text-on-apps-with-both.tost5.ci90-inside\endcsname{1}
\expandafter\def\csname cnt@rev.desc.d7.uiv.tau-T-by-type.p2.center-d1.icon-minus-text-on-apps-with-both.tost5.ci90-inside\endcsname{1}
\expandafter\def\csname pct@rev.desc.d7.uiv.tau-T-by-type.p2.center-d1.icon-minus-text-on-apps-with-both.tost5.ci90-inside\endcsname{100.0}
\expandafter\def\csname ppt@rev.desc.d7.uiv.tau-T-by-type.p2.center-d1.icon-minus-text-on-apps-with-both.tost5.ci90-inside\endcsname{$+$100.0}
\expandafter\def\csname val@rev.desc.d7.uiv.tau-T-by-type.p2.center-d1.icon-minus-text-on-apps-with-both.tost5.equivalent\endcsname{1}
\expandafter\def\csname cnt@rev.desc.d7.uiv.tau-T-by-type.p2.center-d1.icon-minus-text-on-apps-with-both.tost5.equivalent\endcsname{1}
\expandafter\def\csname pct@rev.desc.d7.uiv.tau-T-by-type.p2.center-d1.icon-minus-text-on-apps-with-both.tost5.equivalent\endcsname{100.0}
\expandafter\def\csname ppt@rev.desc.d7.uiv.tau-T-by-type.p2.center-d1.icon-minus-text-on-apps-with-both.tost5.equivalent\endcsname{$+$100.0}
\expandafter\def\csname val@rev.desc.d7.uiv.tau-T-by-type.p2.center-d1.icon-minus-text-on-apps-with-both.tost5.p\endcsname{0.016}
\expandafter\def\csname cnt@rev.desc.d7.uiv.tau-T-by-type.p2.center-d1.icon-minus-text-on-apps-with-both.tost5.p\endcsname{0}
\expandafter\def\csname pct@rev.desc.d7.uiv.tau-T-by-type.p2.center-d1.icon-minus-text-on-apps-with-both.tost5.p\endcsname{1.6}
\expandafter\def\csname ppt@rev.desc.d7.uiv.tau-T-by-type.p2.center-d1.icon-minus-text-on-apps-with-both.tost5.p\endcsname{$+$1.6}
\expandafter\def\csname val@rev.desc.d7.uiv.tau-T-by-type.p2.center-d1.icon.acc-F\endcsname{0.816}
\expandafter\def\csname cnt@rev.desc.d7.uiv.tau-T-by-type.p2.center-d1.icon.acc-F\endcsname{1}
\expandafter\def\csname pct@rev.desc.d7.uiv.tau-T-by-type.p2.center-d1.icon.acc-F\endcsname{81.6}
\expandafter\def\csname ppt@rev.desc.d7.uiv.tau-T-by-type.p2.center-d1.icon.acc-F\endcsname{$+$81.6}
\expandafter\def\csname val@rev.desc.d7.uiv.tau-T-by-type.p2.center-d1.icon.acc-direct\endcsname{0.851}
\expandafter\def\csname cnt@rev.desc.d7.uiv.tau-T-by-type.p2.center-d1.icon.acc-direct\endcsname{1}
\expandafter\def\csname pct@rev.desc.d7.uiv.tau-T-by-type.p2.center-d1.icon.acc-direct\endcsname{85.1}
\expandafter\def\csname ppt@rev.desc.d7.uiv.tau-T-by-type.p2.center-d1.icon.acc-direct\endcsname{$+$85.1}
\expandafter\def\csname val@rev.desc.d7.uiv.tau-T-by-type.p2.center-d1.icon.n-apps\endcsname{23}
\expandafter\def\csname cnt@rev.desc.d7.uiv.tau-T-by-type.p2.center-d1.icon.n-apps\endcsname{23}
\expandafter\def\csname pct@rev.desc.d7.uiv.tau-T-by-type.p2.center-d1.icon.n-apps\endcsname{2300.0}
\expandafter\def\csname ppt@rev.desc.d7.uiv.tau-T-by-type.p2.center-d1.icon.n-apps\endcsname{$+$2300.0}
\expandafter\def\csname val@rev.desc.d7.uiv.tau-T-by-type.p2.center-d1.icon.n-tasks\endcsname{114}
\expandafter\def\csname cnt@rev.desc.d7.uiv.tau-T-by-type.p2.center-d1.icon.n-tasks\endcsname{114}
\expandafter\def\csname pct@rev.desc.d7.uiv.tau-T-by-type.p2.center-d1.icon.n-tasks\endcsname{11400.0}
\expandafter\def\csname ppt@rev.desc.d7.uiv.tau-T-by-type.p2.center-d1.icon.n-tasks\endcsname{$+$11400.0}
\expandafter\def\csname val@rev.desc.d7.uiv.tau-T-by-type.p2.center-d1.icon.tau.ci90.0\endcsname{-0.065}
\expandafter\def\csname cnt@rev.desc.d7.uiv.tau-T-by-type.p2.center-d1.icon.tau.ci90.0\endcsname{0}
\expandafter\def\csname pct@rev.desc.d7.uiv.tau-T-by-type.p2.center-d1.icon.tau.ci90.0\endcsname{$-$6.5}
\expandafter\def\csname ppt@rev.desc.d7.uiv.tau-T-by-type.p2.center-d1.icon.tau.ci90.0\endcsname{$-$6.5}
\expandafter\def\csname val@rev.desc.d7.uiv.tau-T-by-type.p2.center-d1.icon.tau.ci90.1\endcsname{0.009}
\expandafter\def\csname cnt@rev.desc.d7.uiv.tau-T-by-type.p2.center-d1.icon.tau.ci90.1\endcsname{0}
\expandafter\def\csname pct@rev.desc.d7.uiv.tau-T-by-type.p2.center-d1.icon.tau.ci90.1\endcsname{0.9}
\expandafter\def\csname ppt@rev.desc.d7.uiv.tau-T-by-type.p2.center-d1.icon.tau.ci90.1\endcsname{$+$0.9}
\expandafter\def\csname val@rev.desc.d7.uiv.tau-T-by-type.p2.center-d1.icon.tau.ci95-excludes-0\endcsname{0}
\expandafter\def\csname cnt@rev.desc.d7.uiv.tau-T-by-type.p2.center-d1.icon.tau.ci95-excludes-0\endcsname{0}
\expandafter\def\csname pct@rev.desc.d7.uiv.tau-T-by-type.p2.center-d1.icon.tau.ci95-excludes-0\endcsname{0.0}
\expandafter\def\csname ppt@rev.desc.d7.uiv.tau-T-by-type.p2.center-d1.icon.tau.ci95-excludes-0\endcsname{0.0}
\expandafter\def\csname val@rev.desc.d7.uiv.tau-T-by-type.p2.center-d1.icon.tau.ci95.0\endcsname{-0.080}
\expandafter\def\csname cnt@rev.desc.d7.uiv.tau-T-by-type.p2.center-d1.icon.tau.ci95.0\endcsname{0}
\expandafter\def\csname pct@rev.desc.d7.uiv.tau-T-by-type.p2.center-d1.icon.tau.ci95.0\endcsname{$-$8.0}
\expandafter\def\csname ppt@rev.desc.d7.uiv.tau-T-by-type.p2.center-d1.icon.tau.ci95.0\endcsname{$-$8.0}
\expandafter\def\csname val@rev.desc.d7.uiv.tau-T-by-type.p2.center-d1.icon.tau.ci95.1\endcsname{0.012}
\expandafter\def\csname cnt@rev.desc.d7.uiv.tau-T-by-type.p2.center-d1.icon.tau.ci95.1\endcsname{0}
\expandafter\def\csname pct@rev.desc.d7.uiv.tau-T-by-type.p2.center-d1.icon.tau.ci95.1\endcsname{1.2}
\expandafter\def\csname ppt@rev.desc.d7.uiv.tau-T-by-type.p2.center-d1.icon.tau.ci95.1\endcsname{$+$1.2}
\expandafter\def\csname val@rev.desc.d7.uiv.tau-T-by-type.p2.center-d1.icon.tau.mean\endcsname{-0.028}
\expandafter\def\csname cnt@rev.desc.d7.uiv.tau-T-by-type.p2.center-d1.icon.tau.mean\endcsname{0}
\expandafter\def\csname pct@rev.desc.d7.uiv.tau-T-by-type.p2.center-d1.icon.tau.mean\endcsname{$-$2.8}
\expandafter\def\csname ppt@rev.desc.d7.uiv.tau-T-by-type.p2.center-d1.icon.tau.mean\endcsname{$-$2.8}
\expandafter\def\csname val@rev.desc.d7.uiv.tau-T-by-type.p2.center-d1.icon.tau.n-apps\endcsname{23}
\expandafter\def\csname cnt@rev.desc.d7.uiv.tau-T-by-type.p2.center-d1.icon.tau.n-apps\endcsname{23}
\expandafter\def\csname pct@rev.desc.d7.uiv.tau-T-by-type.p2.center-d1.icon.tau.n-apps\endcsname{2300.0}
\expandafter\def\csname ppt@rev.desc.d7.uiv.tau-T-by-type.p2.center-d1.icon.tau.n-apps\endcsname{$+$2300.0}
\expandafter\def\csname val@rev.desc.d7.uiv.tau-T-by-type.p2.center-d1.icon.tau.tost1.ci90-inside\endcsname{0}
\expandafter\def\csname cnt@rev.desc.d7.uiv.tau-T-by-type.p2.center-d1.icon.tau.tost1.ci90-inside\endcsname{0}
\expandafter\def\csname pct@rev.desc.d7.uiv.tau-T-by-type.p2.center-d1.icon.tau.tost1.ci90-inside\endcsname{0.0}
\expandafter\def\csname ppt@rev.desc.d7.uiv.tau-T-by-type.p2.center-d1.icon.tau.tost1.ci90-inside\endcsname{0.0}
\expandafter\def\csname val@rev.desc.d7.uiv.tau-T-by-type.p2.center-d1.icon.tau.tost1.equivalent\endcsname{0}
\expandafter\def\csname cnt@rev.desc.d7.uiv.tau-T-by-type.p2.center-d1.icon.tau.tost1.equivalent\endcsname{0}
\expandafter\def\csname pct@rev.desc.d7.uiv.tau-T-by-type.p2.center-d1.icon.tau.tost1.equivalent\endcsname{0.0}
\expandafter\def\csname ppt@rev.desc.d7.uiv.tau-T-by-type.p2.center-d1.icon.tau.tost1.equivalent\endcsname{0.0}
\expandafter\def\csname val@rev.desc.d7.uiv.tau-T-by-type.p2.center-d1.icon.tau.tost1.p\endcsname{0.778}
\expandafter\def\csname cnt@rev.desc.d7.uiv.tau-T-by-type.p2.center-d1.icon.tau.tost1.p\endcsname{1}
\expandafter\def\csname pct@rev.desc.d7.uiv.tau-T-by-type.p2.center-d1.icon.tau.tost1.p\endcsname{77.8}
\expandafter\def\csname ppt@rev.desc.d7.uiv.tau-T-by-type.p2.center-d1.icon.tau.tost1.p\endcsname{$+$77.8}
\expandafter\def\csname val@rev.desc.d7.uiv.tau-T-by-type.p2.center-d1.icon.tau.tost5.ci90-inside\endcsname{0}
\expandafter\def\csname cnt@rev.desc.d7.uiv.tau-T-by-type.p2.center-d1.icon.tau.tost5.ci90-inside\endcsname{0}
\expandafter\def\csname pct@rev.desc.d7.uiv.tau-T-by-type.p2.center-d1.icon.tau.tost5.ci90-inside\endcsname{0.0}
\expandafter\def\csname ppt@rev.desc.d7.uiv.tau-T-by-type.p2.center-d1.icon.tau.tost5.ci90-inside\endcsname{0.0}
\expandafter\def\csname val@rev.desc.d7.uiv.tau-T-by-type.p2.center-d1.icon.tau.tost5.equivalent\endcsname{0}
\expandafter\def\csname cnt@rev.desc.d7.uiv.tau-T-by-type.p2.center-d1.icon.tau.tost5.equivalent\endcsname{0}
\expandafter\def\csname pct@rev.desc.d7.uiv.tau-T-by-type.p2.center-d1.icon.tau.tost5.equivalent\endcsname{0.0}
\expandafter\def\csname ppt@rev.desc.d7.uiv.tau-T-by-type.p2.center-d1.icon.tau.tost5.equivalent\endcsname{0.0}
\expandafter\def\csname val@rev.desc.d7.uiv.tau-T-by-type.p2.center-d1.icon.tau.tost5.p\endcsname{0.163}
\expandafter\def\csname cnt@rev.desc.d7.uiv.tau-T-by-type.p2.center-d1.icon.tau.tost5.p\endcsname{0}
\expandafter\def\csname pct@rev.desc.d7.uiv.tau-T-by-type.p2.center-d1.icon.tau.tost5.p\endcsname{16.3}
\expandafter\def\csname ppt@rev.desc.d7.uiv.tau-T-by-type.p2.center-d1.icon.tau.tost5.p\endcsname{$+$16.3}
\expandafter\def\csname val@rev.desc.d7.uiv.tau-T-by-type.p2.center-d1.text.acc-F\endcsname{0.800}
\expandafter\def\csname cnt@rev.desc.d7.uiv.tau-T-by-type.p2.center-d1.text.acc-F\endcsname{1}
\expandafter\def\csname pct@rev.desc.d7.uiv.tau-T-by-type.p2.center-d1.text.acc-F\endcsname{80.0}
\expandafter\def\csname ppt@rev.desc.d7.uiv.tau-T-by-type.p2.center-d1.text.acc-F\endcsname{$+$80.0}
\expandafter\def\csname val@rev.desc.d7.uiv.tau-T-by-type.p2.center-d1.text.acc-direct\endcsname{0.800}
\expandafter\def\csname cnt@rev.desc.d7.uiv.tau-T-by-type.p2.center-d1.text.acc-direct\endcsname{1}
\expandafter\def\csname pct@rev.desc.d7.uiv.tau-T-by-type.p2.center-d1.text.acc-direct\endcsname{80.0}
\expandafter\def\csname ppt@rev.desc.d7.uiv.tau-T-by-type.p2.center-d1.text.acc-direct\endcsname{$+$80.0}
\expandafter\def\csname val@rev.desc.d7.uiv.tau-T-by-type.p2.center-d1.text.n-apps\endcsname{15}
\expandafter\def\csname cnt@rev.desc.d7.uiv.tau-T-by-type.p2.center-d1.text.n-apps\endcsname{15}
\expandafter\def\csname pct@rev.desc.d7.uiv.tau-T-by-type.p2.center-d1.text.n-apps\endcsname{1500.0}
\expandafter\def\csname ppt@rev.desc.d7.uiv.tau-T-by-type.p2.center-d1.text.n-apps\endcsname{$+$1500.0}
\expandafter\def\csname val@rev.desc.d7.uiv.tau-T-by-type.p2.center-d1.text.n-tasks\endcsname{30}
\expandafter\def\csname cnt@rev.desc.d7.uiv.tau-T-by-type.p2.center-d1.text.n-tasks\endcsname{30}
\expandafter\def\csname pct@rev.desc.d7.uiv.tau-T-by-type.p2.center-d1.text.n-tasks\endcsname{3000.0}
\expandafter\def\csname ppt@rev.desc.d7.uiv.tau-T-by-type.p2.center-d1.text.n-tasks\endcsname{$+$3000.0}
\expandafter\def\csname val@rev.desc.d7.uiv.tau-T-by-type.p2.center-d1.text.tau.ci90.0\endcsname{0.000}
\expandafter\def\csname cnt@rev.desc.d7.uiv.tau-T-by-type.p2.center-d1.text.tau.ci90.0\endcsname{0}
\expandafter\def\csname pct@rev.desc.d7.uiv.tau-T-by-type.p2.center-d1.text.tau.ci90.0\endcsname{0.0}
\expandafter\def\csname ppt@rev.desc.d7.uiv.tau-T-by-type.p2.center-d1.text.tau.ci90.0\endcsname{0.0}
\expandafter\def\csname val@rev.desc.d7.uiv.tau-T-by-type.p2.center-d1.text.tau.ci90.1\endcsname{0.000}
\expandafter\def\csname cnt@rev.desc.d7.uiv.tau-T-by-type.p2.center-d1.text.tau.ci90.1\endcsname{0}
\expandafter\def\csname pct@rev.desc.d7.uiv.tau-T-by-type.p2.center-d1.text.tau.ci90.1\endcsname{0.0}
\expandafter\def\csname ppt@rev.desc.d7.uiv.tau-T-by-type.p2.center-d1.text.tau.ci90.1\endcsname{0.0}
\expandafter\def\csname val@rev.desc.d7.uiv.tau-T-by-type.p2.center-d1.text.tau.ci95-excludes-0\endcsname{0}
\expandafter\def\csname cnt@rev.desc.d7.uiv.tau-T-by-type.p2.center-d1.text.tau.ci95-excludes-0\endcsname{0}
\expandafter\def\csname pct@rev.desc.d7.uiv.tau-T-by-type.p2.center-d1.text.tau.ci95-excludes-0\endcsname{0.0}
\expandafter\def\csname ppt@rev.desc.d7.uiv.tau-T-by-type.p2.center-d1.text.tau.ci95-excludes-0\endcsname{0.0}
\expandafter\def\csname val@rev.desc.d7.uiv.tau-T-by-type.p2.center-d1.text.tau.ci95.0\endcsname{0.000}
\expandafter\def\csname cnt@rev.desc.d7.uiv.tau-T-by-type.p2.center-d1.text.tau.ci95.0\endcsname{0}
\expandafter\def\csname pct@rev.desc.d7.uiv.tau-T-by-type.p2.center-d1.text.tau.ci95.0\endcsname{0.0}
\expandafter\def\csname ppt@rev.desc.d7.uiv.tau-T-by-type.p2.center-d1.text.tau.ci95.0\endcsname{0.0}
\expandafter\def\csname val@rev.desc.d7.uiv.tau-T-by-type.p2.center-d1.text.tau.ci95.1\endcsname{0.000}
\expandafter\def\csname cnt@rev.desc.d7.uiv.tau-T-by-type.p2.center-d1.text.tau.ci95.1\endcsname{0}
\expandafter\def\csname pct@rev.desc.d7.uiv.tau-T-by-type.p2.center-d1.text.tau.ci95.1\endcsname{0.0}
\expandafter\def\csname ppt@rev.desc.d7.uiv.tau-T-by-type.p2.center-d1.text.tau.ci95.1\endcsname{0.0}
\expandafter\def\csname val@rev.desc.d7.uiv.tau-T-by-type.p2.center-d1.text.tau.mean\endcsname{0.000}
\expandafter\def\csname cnt@rev.desc.d7.uiv.tau-T-by-type.p2.center-d1.text.tau.mean\endcsname{0}
\expandafter\def\csname pct@rev.desc.d7.uiv.tau-T-by-type.p2.center-d1.text.tau.mean\endcsname{0.0}
\expandafter\def\csname ppt@rev.desc.d7.uiv.tau-T-by-type.p2.center-d1.text.tau.mean\endcsname{0.0}
\expandafter\def\csname val@rev.desc.d7.uiv.tau-T-by-type.p2.center-d1.text.tau.n-apps\endcsname{15}
\expandafter\def\csname cnt@rev.desc.d7.uiv.tau-T-by-type.p2.center-d1.text.tau.n-apps\endcsname{15}
\expandafter\def\csname pct@rev.desc.d7.uiv.tau-T-by-type.p2.center-d1.text.tau.n-apps\endcsname{1500.0}
\expandafter\def\csname ppt@rev.desc.d7.uiv.tau-T-by-type.p2.center-d1.text.tau.n-apps\endcsname{$+$1500.0}
\expandafter\def\csname val@rev.desc.d7.uiv.tau-T-by-type.p2.center-d1.text.tau.tost1.ci90-inside\endcsname{1}
\expandafter\def\csname cnt@rev.desc.d7.uiv.tau-T-by-type.p2.center-d1.text.tau.tost1.ci90-inside\endcsname{1}
\expandafter\def\csname pct@rev.desc.d7.uiv.tau-T-by-type.p2.center-d1.text.tau.tost1.ci90-inside\endcsname{100.0}
\expandafter\def\csname ppt@rev.desc.d7.uiv.tau-T-by-type.p2.center-d1.text.tau.tost1.ci90-inside\endcsname{$+$100.0}
\expandafter\def\csname val@rev.desc.d7.uiv.tau-T-by-type.p2.center-d1.text.tau.tost1.equivalent\endcsname{1}
\expandafter\def\csname cnt@rev.desc.d7.uiv.tau-T-by-type.p2.center-d1.text.tau.tost1.equivalent\endcsname{1}
\expandafter\def\csname pct@rev.desc.d7.uiv.tau-T-by-type.p2.center-d1.text.tau.tost1.equivalent\endcsname{100.0}
\expandafter\def\csname ppt@rev.desc.d7.uiv.tau-T-by-type.p2.center-d1.text.tau.tost1.equivalent\endcsname{$+$100.0}
\expandafter\def\csname val@rev.desc.d7.uiv.tau-T-by-type.p2.center-d1.text.tau.tost1.p\endcsname{0.000}
\expandafter\def\csname cnt@rev.desc.d7.uiv.tau-T-by-type.p2.center-d1.text.tau.tost1.p\endcsname{0}
\expandafter\def\csname pct@rev.desc.d7.uiv.tau-T-by-type.p2.center-d1.text.tau.tost1.p\endcsname{0.0}
\expandafter\def\csname ppt@rev.desc.d7.uiv.tau-T-by-type.p2.center-d1.text.tau.tost1.p\endcsname{0.0}
\expandafter\def\csname val@rev.desc.d7.uiv.tau-T-by-type.p2.center-d1.text.tau.tost5.ci90-inside\endcsname{1}
\expandafter\def\csname cnt@rev.desc.d7.uiv.tau-T-by-type.p2.center-d1.text.tau.tost5.ci90-inside\endcsname{1}
\expandafter\def\csname pct@rev.desc.d7.uiv.tau-T-by-type.p2.center-d1.text.tau.tost5.ci90-inside\endcsname{100.0}
\expandafter\def\csname ppt@rev.desc.d7.uiv.tau-T-by-type.p2.center-d1.text.tau.tost5.ci90-inside\endcsname{$+$100.0}
\expandafter\def\csname val@rev.desc.d7.uiv.tau-T-by-type.p2.center-d1.text.tau.tost5.equivalent\endcsname{1}
\expandafter\def\csname cnt@rev.desc.d7.uiv.tau-T-by-type.p2.center-d1.text.tau.tost5.equivalent\endcsname{1}
\expandafter\def\csname pct@rev.desc.d7.uiv.tau-T-by-type.p2.center-d1.text.tau.tost5.equivalent\endcsname{100.0}
\expandafter\def\csname ppt@rev.desc.d7.uiv.tau-T-by-type.p2.center-d1.text.tau.tost5.equivalent\endcsname{$+$100.0}
\expandafter\def\csname val@rev.desc.d7.uiv.tau-T-by-type.p2.center-d1.text.tau.tost5.p\endcsname{0.000}
\expandafter\def\csname cnt@rev.desc.d7.uiv.tau-T-by-type.p2.center-d1.text.tau.tost5.p\endcsname{0}
\expandafter\def\csname pct@rev.desc.d7.uiv.tau-T-by-type.p2.center-d1.text.tau.tost5.p\endcsname{0.0}
\expandafter\def\csname ppt@rev.desc.d7.uiv.tau-T-by-type.p2.center-d1.text.tau.tost5.p\endcsname{0.0}
\expandafter\def\csname val@rev.desc.d7.uiv.tau-T-by-type.p2.context-d1.icon-minus-text-on-apps-with-both.ci90.0\endcsname{0.000}
\expandafter\def\csname cnt@rev.desc.d7.uiv.tau-T-by-type.p2.context-d1.icon-minus-text-on-apps-with-both.ci90.0\endcsname{0}
\expandafter\def\csname pct@rev.desc.d7.uiv.tau-T-by-type.p2.context-d1.icon-minus-text-on-apps-with-both.ci90.0\endcsname{0.0}
\expandafter\def\csname ppt@rev.desc.d7.uiv.tau-T-by-type.p2.context-d1.icon-minus-text-on-apps-with-both.ci90.0\endcsname{0.0}
\expandafter\def\csname val@rev.desc.d7.uiv.tau-T-by-type.p2.context-d1.icon-minus-text-on-apps-with-both.ci90.1\endcsname{0.043}
\expandafter\def\csname cnt@rev.desc.d7.uiv.tau-T-by-type.p2.context-d1.icon-minus-text-on-apps-with-both.ci90.1\endcsname{0}
\expandafter\def\csname pct@rev.desc.d7.uiv.tau-T-by-type.p2.context-d1.icon-minus-text-on-apps-with-both.ci90.1\endcsname{4.3}
\expandafter\def\csname ppt@rev.desc.d7.uiv.tau-T-by-type.p2.context-d1.icon-minus-text-on-apps-with-both.ci90.1\endcsname{$+$4.3}
\expandafter\def\csname val@rev.desc.d7.uiv.tau-T-by-type.p2.context-d1.icon-minus-text-on-apps-with-both.ci95-excludes-0\endcsname{0}
\expandafter\def\csname cnt@rev.desc.d7.uiv.tau-T-by-type.p2.context-d1.icon-minus-text-on-apps-with-both.ci95-excludes-0\endcsname{0}
\expandafter\def\csname pct@rev.desc.d7.uiv.tau-T-by-type.p2.context-d1.icon-minus-text-on-apps-with-both.ci95-excludes-0\endcsname{0.0}
\expandafter\def\csname ppt@rev.desc.d7.uiv.tau-T-by-type.p2.context-d1.icon-minus-text-on-apps-with-both.ci95-excludes-0\endcsname{0.0}
\expandafter\def\csname val@rev.desc.d7.uiv.tau-T-by-type.p2.context-d1.icon-minus-text-on-apps-with-both.ci95.0\endcsname{0.000}
\expandafter\def\csname cnt@rev.desc.d7.uiv.tau-T-by-type.p2.context-d1.icon-minus-text-on-apps-with-both.ci95.0\endcsname{0}
\expandafter\def\csname pct@rev.desc.d7.uiv.tau-T-by-type.p2.context-d1.icon-minus-text-on-apps-with-both.ci95.0\endcsname{0.0}
\expandafter\def\csname ppt@rev.desc.d7.uiv.tau-T-by-type.p2.context-d1.icon-minus-text-on-apps-with-both.ci95.0\endcsname{0.0}
\expandafter\def\csname val@rev.desc.d7.uiv.tau-T-by-type.p2.context-d1.icon-minus-text-on-apps-with-both.ci95.1\endcsname{0.043}
\expandafter\def\csname cnt@rev.desc.d7.uiv.tau-T-by-type.p2.context-d1.icon-minus-text-on-apps-with-both.ci95.1\endcsname{0}
\expandafter\def\csname pct@rev.desc.d7.uiv.tau-T-by-type.p2.context-d1.icon-minus-text-on-apps-with-both.ci95.1\endcsname{4.3}
\expandafter\def\csname ppt@rev.desc.d7.uiv.tau-T-by-type.p2.context-d1.icon-minus-text-on-apps-with-both.ci95.1\endcsname{$+$4.3}
\expandafter\def\csname val@rev.desc.d7.uiv.tau-T-by-type.p2.context-d1.icon-minus-text-on-apps-with-both.mean\endcsname{0.014}
\expandafter\def\csname cnt@rev.desc.d7.uiv.tau-T-by-type.p2.context-d1.icon-minus-text-on-apps-with-both.mean\endcsname{0}
\expandafter\def\csname pct@rev.desc.d7.uiv.tau-T-by-type.p2.context-d1.icon-minus-text-on-apps-with-both.mean\endcsname{1.4}
\expandafter\def\csname ppt@rev.desc.d7.uiv.tau-T-by-type.p2.context-d1.icon-minus-text-on-apps-with-both.mean\endcsname{$+$1.4}
\expandafter\def\csname val@rev.desc.d7.uiv.tau-T-by-type.p2.context-d1.icon-minus-text-on-apps-with-both.n-apps\endcsname{14}
\expandafter\def\csname cnt@rev.desc.d7.uiv.tau-T-by-type.p2.context-d1.icon-minus-text-on-apps-with-both.n-apps\endcsname{14}
\expandafter\def\csname pct@rev.desc.d7.uiv.tau-T-by-type.p2.context-d1.icon-minus-text-on-apps-with-both.n-apps\endcsname{1400.0}
\expandafter\def\csname ppt@rev.desc.d7.uiv.tau-T-by-type.p2.context-d1.icon-minus-text-on-apps-with-both.n-apps\endcsname{$+$1400.0}
\expandafter\def\csname val@rev.desc.d7.uiv.tau-T-by-type.p2.context-d1.icon-minus-text-on-apps-with-both.tost1.ci90-inside\endcsname{0}
\expandafter\def\csname cnt@rev.desc.d7.uiv.tau-T-by-type.p2.context-d1.icon-minus-text-on-apps-with-both.tost1.ci90-inside\endcsname{0}
\expandafter\def\csname pct@rev.desc.d7.uiv.tau-T-by-type.p2.context-d1.icon-minus-text-on-apps-with-both.tost1.ci90-inside\endcsname{0.0}
\expandafter\def\csname ppt@rev.desc.d7.uiv.tau-T-by-type.p2.context-d1.icon-minus-text-on-apps-with-both.tost1.ci90-inside\endcsname{0.0}
\expandafter\def\csname val@rev.desc.d7.uiv.tau-T-by-type.p2.context-d1.icon-minus-text-on-apps-with-both.tost1.equivalent\endcsname{0}
\expandafter\def\csname cnt@rev.desc.d7.uiv.tau-T-by-type.p2.context-d1.icon-minus-text-on-apps-with-both.tost1.equivalent\endcsname{0}
\expandafter\def\csname pct@rev.desc.d7.uiv.tau-T-by-type.p2.context-d1.icon-minus-text-on-apps-with-both.tost1.equivalent\endcsname{0.0}
\expandafter\def\csname ppt@rev.desc.d7.uiv.tau-T-by-type.p2.context-d1.icon-minus-text-on-apps-with-both.tost1.equivalent\endcsname{0.0}
\expandafter\def\csname val@rev.desc.d7.uiv.tau-T-by-type.p2.context-d1.icon-minus-text-on-apps-with-both.tost1.p\endcsname{0.640}
\expandafter\def\csname cnt@rev.desc.d7.uiv.tau-T-by-type.p2.context-d1.icon-minus-text-on-apps-with-both.tost1.p\endcsname{1}
\expandafter\def\csname pct@rev.desc.d7.uiv.tau-T-by-type.p2.context-d1.icon-minus-text-on-apps-with-both.tost1.p\endcsname{64.0}
\expandafter\def\csname ppt@rev.desc.d7.uiv.tau-T-by-type.p2.context-d1.icon-minus-text-on-apps-with-both.tost1.p\endcsname{$+$64.0}
\expandafter\def\csname val@rev.desc.d7.uiv.tau-T-by-type.p2.context-d1.icon-minus-text-on-apps-with-both.tost5.ci90-inside\endcsname{1}
\expandafter\def\csname cnt@rev.desc.d7.uiv.tau-T-by-type.p2.context-d1.icon-minus-text-on-apps-with-both.tost5.ci90-inside\endcsname{1}
\expandafter\def\csname pct@rev.desc.d7.uiv.tau-T-by-type.p2.context-d1.icon-minus-text-on-apps-with-both.tost5.ci90-inside\endcsname{100.0}
\expandafter\def\csname ppt@rev.desc.d7.uiv.tau-T-by-type.p2.context-d1.icon-minus-text-on-apps-with-both.tost5.ci90-inside\endcsname{$+$100.0}
\expandafter\def\csname val@rev.desc.d7.uiv.tau-T-by-type.p2.context-d1.icon-minus-text-on-apps-with-both.tost5.equivalent\endcsname{1}
\expandafter\def\csname cnt@rev.desc.d7.uiv.tau-T-by-type.p2.context-d1.icon-minus-text-on-apps-with-both.tost5.equivalent\endcsname{1}
\expandafter\def\csname pct@rev.desc.d7.uiv.tau-T-by-type.p2.context-d1.icon-minus-text-on-apps-with-both.tost5.equivalent\endcsname{100.0}
\expandafter\def\csname ppt@rev.desc.d7.uiv.tau-T-by-type.p2.context-d1.icon-minus-text-on-apps-with-both.tost5.equivalent\endcsname{$+$100.0}
\expandafter\def\csname val@rev.desc.d7.uiv.tau-T-by-type.p2.context-d1.icon-minus-text-on-apps-with-both.tost5.p\endcsname{0.014}
\expandafter\def\csname cnt@rev.desc.d7.uiv.tau-T-by-type.p2.context-d1.icon-minus-text-on-apps-with-both.tost5.p\endcsname{0}
\expandafter\def\csname pct@rev.desc.d7.uiv.tau-T-by-type.p2.context-d1.icon-minus-text-on-apps-with-both.tost5.p\endcsname{1.4}
\expandafter\def\csname ppt@rev.desc.d7.uiv.tau-T-by-type.p2.context-d1.icon-minus-text-on-apps-with-both.tost5.p\endcsname{$+$1.4}
\expandafter\def\csname val@rev.desc.d7.uiv.tau-T-by-type.p2.context-d1.icon.acc-F\endcsname{0.851}
\expandafter\def\csname cnt@rev.desc.d7.uiv.tau-T-by-type.p2.context-d1.icon.acc-F\endcsname{1}
\expandafter\def\csname pct@rev.desc.d7.uiv.tau-T-by-type.p2.context-d1.icon.acc-F\endcsname{85.1}
\expandafter\def\csname ppt@rev.desc.d7.uiv.tau-T-by-type.p2.context-d1.icon.acc-F\endcsname{$+$85.1}
\expandafter\def\csname val@rev.desc.d7.uiv.tau-T-by-type.p2.context-d1.icon.acc-direct\endcsname{0.851}
\expandafter\def\csname cnt@rev.desc.d7.uiv.tau-T-by-type.p2.context-d1.icon.acc-direct\endcsname{1}
\expandafter\def\csname pct@rev.desc.d7.uiv.tau-T-by-type.p2.context-d1.icon.acc-direct\endcsname{85.1}
\expandafter\def\csname ppt@rev.desc.d7.uiv.tau-T-by-type.p2.context-d1.icon.acc-direct\endcsname{$+$85.1}
\expandafter\def\csname val@rev.desc.d7.uiv.tau-T-by-type.p2.context-d1.icon.n-apps\endcsname{23}
\expandafter\def\csname cnt@rev.desc.d7.uiv.tau-T-by-type.p2.context-d1.icon.n-apps\endcsname{23}
\expandafter\def\csname pct@rev.desc.d7.uiv.tau-T-by-type.p2.context-d1.icon.n-apps\endcsname{2300.0}
\expandafter\def\csname ppt@rev.desc.d7.uiv.tau-T-by-type.p2.context-d1.icon.n-apps\endcsname{$+$2300.0}
\expandafter\def\csname val@rev.desc.d7.uiv.tau-T-by-type.p2.context-d1.icon.n-tasks\endcsname{114}
\expandafter\def\csname cnt@rev.desc.d7.uiv.tau-T-by-type.p2.context-d1.icon.n-tasks\endcsname{114}
\expandafter\def\csname pct@rev.desc.d7.uiv.tau-T-by-type.p2.context-d1.icon.n-tasks\endcsname{11400.0}
\expandafter\def\csname ppt@rev.desc.d7.uiv.tau-T-by-type.p2.context-d1.icon.n-tasks\endcsname{$+$11400.0}
\expandafter\def\csname val@rev.desc.d7.uiv.tau-T-by-type.p2.context-d1.icon.tau.ci90.0\endcsname{-0.022}
\expandafter\def\csname cnt@rev.desc.d7.uiv.tau-T-by-type.p2.context-d1.icon.tau.ci90.0\endcsname{0}
\expandafter\def\csname pct@rev.desc.d7.uiv.tau-T-by-type.p2.context-d1.icon.tau.ci90.0\endcsname{$-$2.2}
\expandafter\def\csname ppt@rev.desc.d7.uiv.tau-T-by-type.p2.context-d1.icon.tau.ci90.0\endcsname{$-$2.2}
\expandafter\def\csname val@rev.desc.d7.uiv.tau-T-by-type.p2.context-d1.icon.tau.ci90.1\endcsname{0.026}
\expandafter\def\csname cnt@rev.desc.d7.uiv.tau-T-by-type.p2.context-d1.icon.tau.ci90.1\endcsname{0}
\expandafter\def\csname pct@rev.desc.d7.uiv.tau-T-by-type.p2.context-d1.icon.tau.ci90.1\endcsname{2.6}
\expandafter\def\csname ppt@rev.desc.d7.uiv.tau-T-by-type.p2.context-d1.icon.tau.ci90.1\endcsname{$+$2.6}
\expandafter\def\csname val@rev.desc.d7.uiv.tau-T-by-type.p2.context-d1.icon.tau.ci95-excludes-0\endcsname{0}
\expandafter\def\csname cnt@rev.desc.d7.uiv.tau-T-by-type.p2.context-d1.icon.tau.ci95-excludes-0\endcsname{0}
\expandafter\def\csname pct@rev.desc.d7.uiv.tau-T-by-type.p2.context-d1.icon.tau.ci95-excludes-0\endcsname{0.0}
\expandafter\def\csname ppt@rev.desc.d7.uiv.tau-T-by-type.p2.context-d1.icon.tau.ci95-excludes-0\endcsname{0.0}
\expandafter\def\csname val@rev.desc.d7.uiv.tau-T-by-type.p2.context-d1.icon.tau.ci95.0\endcsname{-0.029}
\expandafter\def\csname cnt@rev.desc.d7.uiv.tau-T-by-type.p2.context-d1.icon.tau.ci95.0\endcsname{0}
\expandafter\def\csname pct@rev.desc.d7.uiv.tau-T-by-type.p2.context-d1.icon.tau.ci95.0\endcsname{$-$2.9}
\expandafter\def\csname ppt@rev.desc.d7.uiv.tau-T-by-type.p2.context-d1.icon.tau.ci95.0\endcsname{$-$2.9}
\expandafter\def\csname val@rev.desc.d7.uiv.tau-T-by-type.p2.context-d1.icon.tau.ci95.1\endcsname{0.032}
\expandafter\def\csname cnt@rev.desc.d7.uiv.tau-T-by-type.p2.context-d1.icon.tau.ci95.1\endcsname{0}
\expandafter\def\csname pct@rev.desc.d7.uiv.tau-T-by-type.p2.context-d1.icon.tau.ci95.1\endcsname{3.2}
\expandafter\def\csname ppt@rev.desc.d7.uiv.tau-T-by-type.p2.context-d1.icon.tau.ci95.1\endcsname{$+$3.2}
\expandafter\def\csname val@rev.desc.d7.uiv.tau-T-by-type.p2.context-d1.icon.tau.mean\endcsname{0.001}
\expandafter\def\csname cnt@rev.desc.d7.uiv.tau-T-by-type.p2.context-d1.icon.tau.mean\endcsname{0}
\expandafter\def\csname pct@rev.desc.d7.uiv.tau-T-by-type.p2.context-d1.icon.tau.mean\endcsname{0.1}
\expandafter\def\csname ppt@rev.desc.d7.uiv.tau-T-by-type.p2.context-d1.icon.tau.mean\endcsname{$+$0.1}
\expandafter\def\csname val@rev.desc.d7.uiv.tau-T-by-type.p2.context-d1.icon.tau.n-apps\endcsname{23}
\expandafter\def\csname cnt@rev.desc.d7.uiv.tau-T-by-type.p2.context-d1.icon.tau.n-apps\endcsname{23}
\expandafter\def\csname pct@rev.desc.d7.uiv.tau-T-by-type.p2.context-d1.icon.tau.n-apps\endcsname{2300.0}
\expandafter\def\csname ppt@rev.desc.d7.uiv.tau-T-by-type.p2.context-d1.icon.tau.n-apps\endcsname{$+$2300.0}
\expandafter\def\csname val@rev.desc.d7.uiv.tau-T-by-type.p2.context-d1.icon.tau.tost1.ci90-inside\endcsname{0}
\expandafter\def\csname cnt@rev.desc.d7.uiv.tau-T-by-type.p2.context-d1.icon.tau.tost1.ci90-inside\endcsname{0}
\expandafter\def\csname pct@rev.desc.d7.uiv.tau-T-by-type.p2.context-d1.icon.tau.tost1.ci90-inside\endcsname{0.0}
\expandafter\def\csname ppt@rev.desc.d7.uiv.tau-T-by-type.p2.context-d1.icon.tau.tost1.ci90-inside\endcsname{0.0}
\expandafter\def\csname val@rev.desc.d7.uiv.tau-T-by-type.p2.context-d1.icon.tau.tost1.equivalent\endcsname{0}
\expandafter\def\csname cnt@rev.desc.d7.uiv.tau-T-by-type.p2.context-d1.icon.tau.tost1.equivalent\endcsname{0}
\expandafter\def\csname pct@rev.desc.d7.uiv.tau-T-by-type.p2.context-d1.icon.tau.tost1.equivalent\endcsname{0.0}
\expandafter\def\csname ppt@rev.desc.d7.uiv.tau-T-by-type.p2.context-d1.icon.tau.tost1.equivalent\endcsname{0.0}
\expandafter\def\csname val@rev.desc.d7.uiv.tau-T-by-type.p2.context-d1.icon.tau.tost1.p\endcsname{0.288}
\expandafter\def\csname cnt@rev.desc.d7.uiv.tau-T-by-type.p2.context-d1.icon.tau.tost1.p\endcsname{0}
\expandafter\def\csname pct@rev.desc.d7.uiv.tau-T-by-type.p2.context-d1.icon.tau.tost1.p\endcsname{28.8}
\expandafter\def\csname ppt@rev.desc.d7.uiv.tau-T-by-type.p2.context-d1.icon.tau.tost1.p\endcsname{$+$28.8}
\expandafter\def\csname val@rev.desc.d7.uiv.tau-T-by-type.p2.context-d1.icon.tau.tost5.ci90-inside\endcsname{1}
\expandafter\def\csname cnt@rev.desc.d7.uiv.tau-T-by-type.p2.context-d1.icon.tau.tost5.ci90-inside\endcsname{1}
\expandafter\def\csname pct@rev.desc.d7.uiv.tau-T-by-type.p2.context-d1.icon.tau.tost5.ci90-inside\endcsname{100.0}
\expandafter\def\csname ppt@rev.desc.d7.uiv.tau-T-by-type.p2.context-d1.icon.tau.tost5.ci90-inside\endcsname{$+$100.0}
\expandafter\def\csname val@rev.desc.d7.uiv.tau-T-by-type.p2.context-d1.icon.tau.tost5.equivalent\endcsname{1}
\expandafter\def\csname cnt@rev.desc.d7.uiv.tau-T-by-type.p2.context-d1.icon.tau.tost5.equivalent\endcsname{1}
\expandafter\def\csname pct@rev.desc.d7.uiv.tau-T-by-type.p2.context-d1.icon.tau.tost5.equivalent\endcsname{100.0}
\expandafter\def\csname ppt@rev.desc.d7.uiv.tau-T-by-type.p2.context-d1.icon.tau.tost5.equivalent\endcsname{$+$100.0}
\expandafter\def\csname val@rev.desc.d7.uiv.tau-T-by-type.p2.context-d1.icon.tau.tost5.p\endcsname{0.001}
\expandafter\def\csname cnt@rev.desc.d7.uiv.tau-T-by-type.p2.context-d1.icon.tau.tost5.p\endcsname{0}
\expandafter\def\csname pct@rev.desc.d7.uiv.tau-T-by-type.p2.context-d1.icon.tau.tost5.p\endcsname{0.1}
\expandafter\def\csname ppt@rev.desc.d7.uiv.tau-T-by-type.p2.context-d1.icon.tau.tost5.p\endcsname{$+$0.1}
\expandafter\def\csname val@rev.desc.d7.uiv.tau-T-by-type.p2.context-d1.text.acc-F\endcsname{0.800}
\expandafter\def\csname cnt@rev.desc.d7.uiv.tau-T-by-type.p2.context-d1.text.acc-F\endcsname{1}
\expandafter\def\csname pct@rev.desc.d7.uiv.tau-T-by-type.p2.context-d1.text.acc-F\endcsname{80.0}
\expandafter\def\csname ppt@rev.desc.d7.uiv.tau-T-by-type.p2.context-d1.text.acc-F\endcsname{$+$80.0}
\expandafter\def\csname val@rev.desc.d7.uiv.tau-T-by-type.p2.context-d1.text.acc-direct\endcsname{0.800}
\expandafter\def\csname cnt@rev.desc.d7.uiv.tau-T-by-type.p2.context-d1.text.acc-direct\endcsname{1}
\expandafter\def\csname pct@rev.desc.d7.uiv.tau-T-by-type.p2.context-d1.text.acc-direct\endcsname{80.0}
\expandafter\def\csname ppt@rev.desc.d7.uiv.tau-T-by-type.p2.context-d1.text.acc-direct\endcsname{$+$80.0}
\expandafter\def\csname val@rev.desc.d7.uiv.tau-T-by-type.p2.context-d1.text.n-apps\endcsname{15}
\expandafter\def\csname cnt@rev.desc.d7.uiv.tau-T-by-type.p2.context-d1.text.n-apps\endcsname{15}
\expandafter\def\csname pct@rev.desc.d7.uiv.tau-T-by-type.p2.context-d1.text.n-apps\endcsname{1500.0}
\expandafter\def\csname ppt@rev.desc.d7.uiv.tau-T-by-type.p2.context-d1.text.n-apps\endcsname{$+$1500.0}
\expandafter\def\csname val@rev.desc.d7.uiv.tau-T-by-type.p2.context-d1.text.n-tasks\endcsname{30}
\expandafter\def\csname cnt@rev.desc.d7.uiv.tau-T-by-type.p2.context-d1.text.n-tasks\endcsname{30}
\expandafter\def\csname pct@rev.desc.d7.uiv.tau-T-by-type.p2.context-d1.text.n-tasks\endcsname{3000.0}
\expandafter\def\csname ppt@rev.desc.d7.uiv.tau-T-by-type.p2.context-d1.text.n-tasks\endcsname{$+$3000.0}
\expandafter\def\csname val@rev.desc.d7.uiv.tau-T-by-type.p2.context-d1.text.tau.ci90.0\endcsname{0.000}
\expandafter\def\csname cnt@rev.desc.d7.uiv.tau-T-by-type.p2.context-d1.text.tau.ci90.0\endcsname{0}
\expandafter\def\csname pct@rev.desc.d7.uiv.tau-T-by-type.p2.context-d1.text.tau.ci90.0\endcsname{0.0}
\expandafter\def\csname ppt@rev.desc.d7.uiv.tau-T-by-type.p2.context-d1.text.tau.ci90.0\endcsname{0.0}
\expandafter\def\csname val@rev.desc.d7.uiv.tau-T-by-type.p2.context-d1.text.tau.ci90.1\endcsname{0.000}
\expandafter\def\csname cnt@rev.desc.d7.uiv.tau-T-by-type.p2.context-d1.text.tau.ci90.1\endcsname{0}
\expandafter\def\csname pct@rev.desc.d7.uiv.tau-T-by-type.p2.context-d1.text.tau.ci90.1\endcsname{0.0}
\expandafter\def\csname ppt@rev.desc.d7.uiv.tau-T-by-type.p2.context-d1.text.tau.ci90.1\endcsname{0.0}
\expandafter\def\csname val@rev.desc.d7.uiv.tau-T-by-type.p2.context-d1.text.tau.ci95-excludes-0\endcsname{0}
\expandafter\def\csname cnt@rev.desc.d7.uiv.tau-T-by-type.p2.context-d1.text.tau.ci95-excludes-0\endcsname{0}
\expandafter\def\csname pct@rev.desc.d7.uiv.tau-T-by-type.p2.context-d1.text.tau.ci95-excludes-0\endcsname{0.0}
\expandafter\def\csname ppt@rev.desc.d7.uiv.tau-T-by-type.p2.context-d1.text.tau.ci95-excludes-0\endcsname{0.0}
\expandafter\def\csname val@rev.desc.d7.uiv.tau-T-by-type.p2.context-d1.text.tau.ci95.0\endcsname{0.000}
\expandafter\def\csname cnt@rev.desc.d7.uiv.tau-T-by-type.p2.context-d1.text.tau.ci95.0\endcsname{0}
\expandafter\def\csname pct@rev.desc.d7.uiv.tau-T-by-type.p2.context-d1.text.tau.ci95.0\endcsname{0.0}
\expandafter\def\csname ppt@rev.desc.d7.uiv.tau-T-by-type.p2.context-d1.text.tau.ci95.0\endcsname{0.0}
\expandafter\def\csname val@rev.desc.d7.uiv.tau-T-by-type.p2.context-d1.text.tau.ci95.1\endcsname{0.000}
\expandafter\def\csname cnt@rev.desc.d7.uiv.tau-T-by-type.p2.context-d1.text.tau.ci95.1\endcsname{0}
\expandafter\def\csname pct@rev.desc.d7.uiv.tau-T-by-type.p2.context-d1.text.tau.ci95.1\endcsname{0.0}
\expandafter\def\csname ppt@rev.desc.d7.uiv.tau-T-by-type.p2.context-d1.text.tau.ci95.1\endcsname{0.0}
\expandafter\def\csname val@rev.desc.d7.uiv.tau-T-by-type.p2.context-d1.text.tau.mean\endcsname{0.000}
\expandafter\def\csname cnt@rev.desc.d7.uiv.tau-T-by-type.p2.context-d1.text.tau.mean\endcsname{0}
\expandafter\def\csname pct@rev.desc.d7.uiv.tau-T-by-type.p2.context-d1.text.tau.mean\endcsname{0.0}
\expandafter\def\csname ppt@rev.desc.d7.uiv.tau-T-by-type.p2.context-d1.text.tau.mean\endcsname{0.0}
\expandafter\def\csname val@rev.desc.d7.uiv.tau-T-by-type.p2.context-d1.text.tau.n-apps\endcsname{15}
\expandafter\def\csname cnt@rev.desc.d7.uiv.tau-T-by-type.p2.context-d1.text.tau.n-apps\endcsname{15}
\expandafter\def\csname pct@rev.desc.d7.uiv.tau-T-by-type.p2.context-d1.text.tau.n-apps\endcsname{1500.0}
\expandafter\def\csname ppt@rev.desc.d7.uiv.tau-T-by-type.p2.context-d1.text.tau.n-apps\endcsname{$+$1500.0}
\expandafter\def\csname val@rev.desc.d7.uiv.tau-T-by-type.p2.context-d1.text.tau.tost1.ci90-inside\endcsname{1}
\expandafter\def\csname cnt@rev.desc.d7.uiv.tau-T-by-type.p2.context-d1.text.tau.tost1.ci90-inside\endcsname{1}
\expandafter\def\csname pct@rev.desc.d7.uiv.tau-T-by-type.p2.context-d1.text.tau.tost1.ci90-inside\endcsname{100.0}
\expandafter\def\csname ppt@rev.desc.d7.uiv.tau-T-by-type.p2.context-d1.text.tau.tost1.ci90-inside\endcsname{$+$100.0}
\expandafter\def\csname val@rev.desc.d7.uiv.tau-T-by-type.p2.context-d1.text.tau.tost1.equivalent\endcsname{1}
\expandafter\def\csname cnt@rev.desc.d7.uiv.tau-T-by-type.p2.context-d1.text.tau.tost1.equivalent\endcsname{1}
\expandafter\def\csname pct@rev.desc.d7.uiv.tau-T-by-type.p2.context-d1.text.tau.tost1.equivalent\endcsname{100.0}
\expandafter\def\csname ppt@rev.desc.d7.uiv.tau-T-by-type.p2.context-d1.text.tau.tost1.equivalent\endcsname{$+$100.0}
\expandafter\def\csname val@rev.desc.d7.uiv.tau-T-by-type.p2.context-d1.text.tau.tost1.p\endcsname{0.000}
\expandafter\def\csname cnt@rev.desc.d7.uiv.tau-T-by-type.p2.context-d1.text.tau.tost1.p\endcsname{0}
\expandafter\def\csname pct@rev.desc.d7.uiv.tau-T-by-type.p2.context-d1.text.tau.tost1.p\endcsname{0.0}
\expandafter\def\csname ppt@rev.desc.d7.uiv.tau-T-by-type.p2.context-d1.text.tau.tost1.p\endcsname{0.0}
\expandafter\def\csname val@rev.desc.d7.uiv.tau-T-by-type.p2.context-d1.text.tau.tost5.ci90-inside\endcsname{1}
\expandafter\def\csname cnt@rev.desc.d7.uiv.tau-T-by-type.p2.context-d1.text.tau.tost5.ci90-inside\endcsname{1}
\expandafter\def\csname pct@rev.desc.d7.uiv.tau-T-by-type.p2.context-d1.text.tau.tost5.ci90-inside\endcsname{100.0}
\expandafter\def\csname ppt@rev.desc.d7.uiv.tau-T-by-type.p2.context-d1.text.tau.tost5.ci90-inside\endcsname{$+$100.0}
\expandafter\def\csname val@rev.desc.d7.uiv.tau-T-by-type.p2.context-d1.text.tau.tost5.equivalent\endcsname{1}
\expandafter\def\csname cnt@rev.desc.d7.uiv.tau-T-by-type.p2.context-d1.text.tau.tost5.equivalent\endcsname{1}
\expandafter\def\csname pct@rev.desc.d7.uiv.tau-T-by-type.p2.context-d1.text.tau.tost5.equivalent\endcsname{100.0}
\expandafter\def\csname ppt@rev.desc.d7.uiv.tau-T-by-type.p2.context-d1.text.tau.tost5.equivalent\endcsname{$+$100.0}
\expandafter\def\csname val@rev.desc.d7.uiv.tau-T-by-type.p2.context-d1.text.tau.tost5.p\endcsname{0.000}
\expandafter\def\csname cnt@rev.desc.d7.uiv.tau-T-by-type.p2.context-d1.text.tau.tost5.p\endcsname{0}
\expandafter\def\csname pct@rev.desc.d7.uiv.tau-T-by-type.p2.context-d1.text.tau.tost5.p\endcsname{0.0}
\expandafter\def\csname ppt@rev.desc.d7.uiv.tau-T-by-type.p2.context-d1.text.tau.tost5.p\endcsname{0.0}
\expandafter\def\csname val@rev.desc.d7.uiv.tau-T-by-type.p2.dual-d2.icon-minus-text-on-apps-with-both.ci90.0\endcsname{-0.314}
\expandafter\def\csname cnt@rev.desc.d7.uiv.tau-T-by-type.p2.dual-d2.icon-minus-text-on-apps-with-both.ci90.0\endcsname{0}
\expandafter\def\csname pct@rev.desc.d7.uiv.tau-T-by-type.p2.dual-d2.icon-minus-text-on-apps-with-both.ci90.0\endcsname{$-$31.4}
\expandafter\def\csname ppt@rev.desc.d7.uiv.tau-T-by-type.p2.dual-d2.icon-minus-text-on-apps-with-both.ci90.0\endcsname{$-$31.4}
\expandafter\def\csname val@rev.desc.d7.uiv.tau-T-by-type.p2.dual-d2.icon-minus-text-on-apps-with-both.ci90.1\endcsname{0.000}
\expandafter\def\csname cnt@rev.desc.d7.uiv.tau-T-by-type.p2.dual-d2.icon-minus-text-on-apps-with-both.ci90.1\endcsname{0}
\expandafter\def\csname pct@rev.desc.d7.uiv.tau-T-by-type.p2.dual-d2.icon-minus-text-on-apps-with-both.ci90.1\endcsname{0.0}
\expandafter\def\csname ppt@rev.desc.d7.uiv.tau-T-by-type.p2.dual-d2.icon-minus-text-on-apps-with-both.ci90.1\endcsname{0.0}
\expandafter\def\csname val@rev.desc.d7.uiv.tau-T-by-type.p2.dual-d2.icon-minus-text-on-apps-with-both.ci95-excludes-0\endcsname{0}
\expandafter\def\csname cnt@rev.desc.d7.uiv.tau-T-by-type.p2.dual-d2.icon-minus-text-on-apps-with-both.ci95-excludes-0\endcsname{0}
\expandafter\def\csname pct@rev.desc.d7.uiv.tau-T-by-type.p2.dual-d2.icon-minus-text-on-apps-with-both.ci95-excludes-0\endcsname{0.0}
\expandafter\def\csname ppt@rev.desc.d7.uiv.tau-T-by-type.p2.dual-d2.icon-minus-text-on-apps-with-both.ci95-excludes-0\endcsname{0.0}
\expandafter\def\csname val@rev.desc.d7.uiv.tau-T-by-type.p2.dual-d2.icon-minus-text-on-apps-with-both.ci95.0\endcsname{-0.329}
\expandafter\def\csname cnt@rev.desc.d7.uiv.tau-T-by-type.p2.dual-d2.icon-minus-text-on-apps-with-both.ci95.0\endcsname{0}
\expandafter\def\csname pct@rev.desc.d7.uiv.tau-T-by-type.p2.dual-d2.icon-minus-text-on-apps-with-both.ci95.0\endcsname{$-$32.9}
\expandafter\def\csname ppt@rev.desc.d7.uiv.tau-T-by-type.p2.dual-d2.icon-minus-text-on-apps-with-both.ci95.0\endcsname{$-$32.9}
\expandafter\def\csname val@rev.desc.d7.uiv.tau-T-by-type.p2.dual-d2.icon-minus-text-on-apps-with-both.ci95.1\endcsname{0.000}
\expandafter\def\csname cnt@rev.desc.d7.uiv.tau-T-by-type.p2.dual-d2.icon-minus-text-on-apps-with-both.ci95.1\endcsname{0}
\expandafter\def\csname pct@rev.desc.d7.uiv.tau-T-by-type.p2.dual-d2.icon-minus-text-on-apps-with-both.ci95.1\endcsname{0.0}
\expandafter\def\csname ppt@rev.desc.d7.uiv.tau-T-by-type.p2.dual-d2.icon-minus-text-on-apps-with-both.ci95.1\endcsname{0.0}
\expandafter\def\csname val@rev.desc.d7.uiv.tau-T-by-type.p2.dual-d2.icon-minus-text-on-apps-with-both.mean\endcsname{-0.114}
\expandafter\def\csname cnt@rev.desc.d7.uiv.tau-T-by-type.p2.dual-d2.icon-minus-text-on-apps-with-both.mean\endcsname{0}
\expandafter\def\csname pct@rev.desc.d7.uiv.tau-T-by-type.p2.dual-d2.icon-minus-text-on-apps-with-both.mean\endcsname{$-$11.4}
\expandafter\def\csname ppt@rev.desc.d7.uiv.tau-T-by-type.p2.dual-d2.icon-minus-text-on-apps-with-both.mean\endcsname{$-$11.4}
\expandafter\def\csname val@rev.desc.d7.uiv.tau-T-by-type.p2.dual-d2.icon-minus-text-on-apps-with-both.n-apps\endcsname{14}
\expandafter\def\csname cnt@rev.desc.d7.uiv.tau-T-by-type.p2.dual-d2.icon-minus-text-on-apps-with-both.n-apps\endcsname{14}
\expandafter\def\csname pct@rev.desc.d7.uiv.tau-T-by-type.p2.dual-d2.icon-minus-text-on-apps-with-both.n-apps\endcsname{1400.0}
\expandafter\def\csname ppt@rev.desc.d7.uiv.tau-T-by-type.p2.dual-d2.icon-minus-text-on-apps-with-both.n-apps\endcsname{$+$1400.0}
\expandafter\def\csname val@rev.desc.d7.uiv.tau-T-by-type.p2.dual-d2.icon-minus-text-on-apps-with-both.tost1.ci90-inside\endcsname{0}
\expandafter\def\csname cnt@rev.desc.d7.uiv.tau-T-by-type.p2.dual-d2.icon-minus-text-on-apps-with-both.tost1.ci90-inside\endcsname{0}
\expandafter\def\csname pct@rev.desc.d7.uiv.tau-T-by-type.p2.dual-d2.icon-minus-text-on-apps-with-both.tost1.ci90-inside\endcsname{0.0}
\expandafter\def\csname ppt@rev.desc.d7.uiv.tau-T-by-type.p2.dual-d2.icon-minus-text-on-apps-with-both.tost1.ci90-inside\endcsname{0.0}
\expandafter\def\csname val@rev.desc.d7.uiv.tau-T-by-type.p2.dual-d2.icon-minus-text-on-apps-with-both.tost1.equivalent\endcsname{0}
\expandafter\def\csname cnt@rev.desc.d7.uiv.tau-T-by-type.p2.dual-d2.icon-minus-text-on-apps-with-both.tost1.equivalent\endcsname{0}
\expandafter\def\csname pct@rev.desc.d7.uiv.tau-T-by-type.p2.dual-d2.icon-minus-text-on-apps-with-both.tost1.equivalent\endcsname{0.0}
\expandafter\def\csname ppt@rev.desc.d7.uiv.tau-T-by-type.p2.dual-d2.icon-minus-text-on-apps-with-both.tost1.equivalent\endcsname{0.0}
\expandafter\def\csname val@rev.desc.d7.uiv.tau-T-by-type.p2.dual-d2.icon-minus-text-on-apps-with-both.tost1.p\endcsname{0.886}
\expandafter\def\csname cnt@rev.desc.d7.uiv.tau-T-by-type.p2.dual-d2.icon-minus-text-on-apps-with-both.tost1.p\endcsname{1}
\expandafter\def\csname pct@rev.desc.d7.uiv.tau-T-by-type.p2.dual-d2.icon-minus-text-on-apps-with-both.tost1.p\endcsname{88.6}
\expandafter\def\csname ppt@rev.desc.d7.uiv.tau-T-by-type.p2.dual-d2.icon-minus-text-on-apps-with-both.tost1.p\endcsname{$+$88.6}
\expandafter\def\csname val@rev.desc.d7.uiv.tau-T-by-type.p2.dual-d2.icon-minus-text-on-apps-with-both.tost5.ci90-inside\endcsname{0}
\expandafter\def\csname cnt@rev.desc.d7.uiv.tau-T-by-type.p2.dual-d2.icon-minus-text-on-apps-with-both.tost5.ci90-inside\endcsname{0}
\expandafter\def\csname pct@rev.desc.d7.uiv.tau-T-by-type.p2.dual-d2.icon-minus-text-on-apps-with-both.tost5.ci90-inside\endcsname{0.0}
\expandafter\def\csname ppt@rev.desc.d7.uiv.tau-T-by-type.p2.dual-d2.icon-minus-text-on-apps-with-both.tost5.ci90-inside\endcsname{0.0}
\expandafter\def\csname val@rev.desc.d7.uiv.tau-T-by-type.p2.dual-d2.icon-minus-text-on-apps-with-both.tost5.equivalent\endcsname{0}
\expandafter\def\csname cnt@rev.desc.d7.uiv.tau-T-by-type.p2.dual-d2.icon-minus-text-on-apps-with-both.tost5.equivalent\endcsname{0}
\expandafter\def\csname pct@rev.desc.d7.uiv.tau-T-by-type.p2.dual-d2.icon-minus-text-on-apps-with-both.tost5.equivalent\endcsname{0.0}
\expandafter\def\csname ppt@rev.desc.d7.uiv.tau-T-by-type.p2.dual-d2.icon-minus-text-on-apps-with-both.tost5.equivalent\endcsname{0.0}
\expandafter\def\csname val@rev.desc.d7.uiv.tau-T-by-type.p2.dual-d2.icon-minus-text-on-apps-with-both.tost5.p\endcsname{0.664}
\expandafter\def\csname cnt@rev.desc.d7.uiv.tau-T-by-type.p2.dual-d2.icon-minus-text-on-apps-with-both.tost5.p\endcsname{1}
\expandafter\def\csname pct@rev.desc.d7.uiv.tau-T-by-type.p2.dual-d2.icon-minus-text-on-apps-with-both.tost5.p\endcsname{66.4}
\expandafter\def\csname ppt@rev.desc.d7.uiv.tau-T-by-type.p2.dual-d2.icon-minus-text-on-apps-with-both.tost5.p\endcsname{$+$66.4}
\expandafter\def\csname val@rev.desc.d7.uiv.tau-T-by-type.p2.dual-d2.icon.acc-F\endcsname{0.763}
\expandafter\def\csname cnt@rev.desc.d7.uiv.tau-T-by-type.p2.dual-d2.icon.acc-F\endcsname{1}
\expandafter\def\csname pct@rev.desc.d7.uiv.tau-T-by-type.p2.dual-d2.icon.acc-F\endcsname{76.3}
\expandafter\def\csname ppt@rev.desc.d7.uiv.tau-T-by-type.p2.dual-d2.icon.acc-F\endcsname{$+$76.3}
\expandafter\def\csname val@rev.desc.d7.uiv.tau-T-by-type.p2.dual-d2.icon.acc-direct\endcsname{0.851}
\expandafter\def\csname cnt@rev.desc.d7.uiv.tau-T-by-type.p2.dual-d2.icon.acc-direct\endcsname{1}
\expandafter\def\csname pct@rev.desc.d7.uiv.tau-T-by-type.p2.dual-d2.icon.acc-direct\endcsname{85.1}
\expandafter\def\csname ppt@rev.desc.d7.uiv.tau-T-by-type.p2.dual-d2.icon.acc-direct\endcsname{$+$85.1}
\expandafter\def\csname val@rev.desc.d7.uiv.tau-T-by-type.p2.dual-d2.icon.n-apps\endcsname{23}
\expandafter\def\csname cnt@rev.desc.d7.uiv.tau-T-by-type.p2.dual-d2.icon.n-apps\endcsname{23}
\expandafter\def\csname pct@rev.desc.d7.uiv.tau-T-by-type.p2.dual-d2.icon.n-apps\endcsname{2300.0}
\expandafter\def\csname ppt@rev.desc.d7.uiv.tau-T-by-type.p2.dual-d2.icon.n-apps\endcsname{$+$2300.0}
\expandafter\def\csname val@rev.desc.d7.uiv.tau-T-by-type.p2.dual-d2.icon.n-tasks\endcsname{114}
\expandafter\def\csname cnt@rev.desc.d7.uiv.tau-T-by-type.p2.dual-d2.icon.n-tasks\endcsname{114}
\expandafter\def\csname pct@rev.desc.d7.uiv.tau-T-by-type.p2.dual-d2.icon.n-tasks\endcsname{11400.0}
\expandafter\def\csname ppt@rev.desc.d7.uiv.tau-T-by-type.p2.dual-d2.icon.n-tasks\endcsname{$+$11400.0}
\expandafter\def\csname val@rev.desc.d7.uiv.tau-T-by-type.p2.dual-d2.icon.tau.ci90.0\endcsname{-0.138}
\expandafter\def\csname cnt@rev.desc.d7.uiv.tau-T-by-type.p2.dual-d2.icon.tau.ci90.0\endcsname{0}
\expandafter\def\csname pct@rev.desc.d7.uiv.tau-T-by-type.p2.dual-d2.icon.tau.ci90.0\endcsname{$-$13.8}
\expandafter\def\csname ppt@rev.desc.d7.uiv.tau-T-by-type.p2.dual-d2.icon.tau.ci90.0\endcsname{$-$13.8}
\expandafter\def\csname val@rev.desc.d7.uiv.tau-T-by-type.p2.dual-d2.icon.tau.ci90.1\endcsname{-0.022}
\expandafter\def\csname cnt@rev.desc.d7.uiv.tau-T-by-type.p2.dual-d2.icon.tau.ci90.1\endcsname{0}
\expandafter\def\csname pct@rev.desc.d7.uiv.tau-T-by-type.p2.dual-d2.icon.tau.ci90.1\endcsname{$-$2.2}
\expandafter\def\csname ppt@rev.desc.d7.uiv.tau-T-by-type.p2.dual-d2.icon.tau.ci90.1\endcsname{$-$2.2}
\expandafter\def\csname val@rev.desc.d7.uiv.tau-T-by-type.p2.dual-d2.icon.tau.ci95-excludes-0\endcsname{1}
\expandafter\def\csname cnt@rev.desc.d7.uiv.tau-T-by-type.p2.dual-d2.icon.tau.ci95-excludes-0\endcsname{1}
\expandafter\def\csname pct@rev.desc.d7.uiv.tau-T-by-type.p2.dual-d2.icon.tau.ci95-excludes-0\endcsname{100.0}
\expandafter\def\csname ppt@rev.desc.d7.uiv.tau-T-by-type.p2.dual-d2.icon.tau.ci95-excludes-0\endcsname{$+$100.0}
\expandafter\def\csname val@rev.desc.d7.uiv.tau-T-by-type.p2.dual-d2.icon.tau.ci95.0\endcsname{-0.152}
\expandafter\def\csname cnt@rev.desc.d7.uiv.tau-T-by-type.p2.dual-d2.icon.tau.ci95.0\endcsname{0}
\expandafter\def\csname pct@rev.desc.d7.uiv.tau-T-by-type.p2.dual-d2.icon.tau.ci95.0\endcsname{$-$15.2}
\expandafter\def\csname ppt@rev.desc.d7.uiv.tau-T-by-type.p2.dual-d2.icon.tau.ci95.0\endcsname{$-$15.2}
\expandafter\def\csname val@rev.desc.d7.uiv.tau-T-by-type.p2.dual-d2.icon.tau.ci95.1\endcsname{-0.012}
\expandafter\def\csname cnt@rev.desc.d7.uiv.tau-T-by-type.p2.dual-d2.icon.tau.ci95.1\endcsname{0}
\expandafter\def\csname pct@rev.desc.d7.uiv.tau-T-by-type.p2.dual-d2.icon.tau.ci95.1\endcsname{$-$1.2}
\expandafter\def\csname ppt@rev.desc.d7.uiv.tau-T-by-type.p2.dual-d2.icon.tau.ci95.1\endcsname{$-$1.2}
\expandafter\def\csname val@rev.desc.d7.uiv.tau-T-by-type.p2.dual-d2.icon.tau.mean\endcsname{-0.077}
\expandafter\def\csname cnt@rev.desc.d7.uiv.tau-T-by-type.p2.dual-d2.icon.tau.mean\endcsname{0}
\expandafter\def\csname pct@rev.desc.d7.uiv.tau-T-by-type.p2.dual-d2.icon.tau.mean\endcsname{$-$7.7}
\expandafter\def\csname ppt@rev.desc.d7.uiv.tau-T-by-type.p2.dual-d2.icon.tau.mean\endcsname{$-$7.7}
\expandafter\def\csname val@rev.desc.d7.uiv.tau-T-by-type.p2.dual-d2.icon.tau.n-apps\endcsname{23}
\expandafter\def\csname cnt@rev.desc.d7.uiv.tau-T-by-type.p2.dual-d2.icon.tau.n-apps\endcsname{23}
\expandafter\def\csname pct@rev.desc.d7.uiv.tau-T-by-type.p2.dual-d2.icon.tau.n-apps\endcsname{2300.0}
\expandafter\def\csname ppt@rev.desc.d7.uiv.tau-T-by-type.p2.dual-d2.icon.tau.n-apps\endcsname{$+$2300.0}
\expandafter\def\csname val@rev.desc.d7.uiv.tau-T-by-type.p2.dual-d2.icon.tau.tost1.ci90-inside\endcsname{0}
\expandafter\def\csname cnt@rev.desc.d7.uiv.tau-T-by-type.p2.dual-d2.icon.tau.tost1.ci90-inside\endcsname{0}
\expandafter\def\csname pct@rev.desc.d7.uiv.tau-T-by-type.p2.dual-d2.icon.tau.tost1.ci90-inside\endcsname{0.0}
\expandafter\def\csname ppt@rev.desc.d7.uiv.tau-T-by-type.p2.dual-d2.icon.tau.tost1.ci90-inside\endcsname{0.0}
\expandafter\def\csname val@rev.desc.d7.uiv.tau-T-by-type.p2.dual-d2.icon.tau.tost1.equivalent\endcsname{0}
\expandafter\def\csname cnt@rev.desc.d7.uiv.tau-T-by-type.p2.dual-d2.icon.tau.tost1.equivalent\endcsname{0}
\expandafter\def\csname pct@rev.desc.d7.uiv.tau-T-by-type.p2.dual-d2.icon.tau.tost1.equivalent\endcsname{0.0}
\expandafter\def\csname ppt@rev.desc.d7.uiv.tau-T-by-type.p2.dual-d2.icon.tau.tost1.equivalent\endcsname{0.0}
\expandafter\def\csname val@rev.desc.d7.uiv.tau-T-by-type.p2.dual-d2.icon.tau.tost1.p\endcsname{0.978}
\expandafter\def\csname cnt@rev.desc.d7.uiv.tau-T-by-type.p2.dual-d2.icon.tau.tost1.p\endcsname{1}
\expandafter\def\csname pct@rev.desc.d7.uiv.tau-T-by-type.p2.dual-d2.icon.tau.tost1.p\endcsname{97.8}
\expandafter\def\csname ppt@rev.desc.d7.uiv.tau-T-by-type.p2.dual-d2.icon.tau.tost1.p\endcsname{$+$97.8}
\expandafter\def\csname val@rev.desc.d7.uiv.tau-T-by-type.p2.dual-d2.icon.tau.tost5.ci90-inside\endcsname{0}
\expandafter\def\csname cnt@rev.desc.d7.uiv.tau-T-by-type.p2.dual-d2.icon.tau.tost5.ci90-inside\endcsname{0}
\expandafter\def\csname pct@rev.desc.d7.uiv.tau-T-by-type.p2.dual-d2.icon.tau.tost5.ci90-inside\endcsname{0.0}
\expandafter\def\csname ppt@rev.desc.d7.uiv.tau-T-by-type.p2.dual-d2.icon.tau.tost5.ci90-inside\endcsname{0.0}
\expandafter\def\csname val@rev.desc.d7.uiv.tau-T-by-type.p2.dual-d2.icon.tau.tost5.equivalent\endcsname{0}
\expandafter\def\csname cnt@rev.desc.d7.uiv.tau-T-by-type.p2.dual-d2.icon.tau.tost5.equivalent\endcsname{0}
\expandafter\def\csname pct@rev.desc.d7.uiv.tau-T-by-type.p2.dual-d2.icon.tau.tost5.equivalent\endcsname{0.0}
\expandafter\def\csname ppt@rev.desc.d7.uiv.tau-T-by-type.p2.dual-d2.icon.tau.tost5.equivalent\endcsname{0.0}
\expandafter\def\csname val@rev.desc.d7.uiv.tau-T-by-type.p2.dual-d2.icon.tau.tost5.p\endcsname{0.782}
\expandafter\def\csname cnt@rev.desc.d7.uiv.tau-T-by-type.p2.dual-d2.icon.tau.tost5.p\endcsname{1}
\expandafter\def\csname pct@rev.desc.d7.uiv.tau-T-by-type.p2.dual-d2.icon.tau.tost5.p\endcsname{78.2}
\expandafter\def\csname ppt@rev.desc.d7.uiv.tau-T-by-type.p2.dual-d2.icon.tau.tost5.p\endcsname{$+$78.2}
\expandafter\def\csname val@rev.desc.d7.uiv.tau-T-by-type.p2.dual-d2.text.acc-F\endcsname{0.833}
\expandafter\def\csname cnt@rev.desc.d7.uiv.tau-T-by-type.p2.dual-d2.text.acc-F\endcsname{1}
\expandafter\def\csname pct@rev.desc.d7.uiv.tau-T-by-type.p2.dual-d2.text.acc-F\endcsname{83.3}
\expandafter\def\csname ppt@rev.desc.d7.uiv.tau-T-by-type.p2.dual-d2.text.acc-F\endcsname{$+$83.3}
\expandafter\def\csname val@rev.desc.d7.uiv.tau-T-by-type.p2.dual-d2.text.acc-direct\endcsname{0.800}
\expandafter\def\csname cnt@rev.desc.d7.uiv.tau-T-by-type.p2.dual-d2.text.acc-direct\endcsname{1}
\expandafter\def\csname pct@rev.desc.d7.uiv.tau-T-by-type.p2.dual-d2.text.acc-direct\endcsname{80.0}
\expandafter\def\csname ppt@rev.desc.d7.uiv.tau-T-by-type.p2.dual-d2.text.acc-direct\endcsname{$+$80.0}
\expandafter\def\csname val@rev.desc.d7.uiv.tau-T-by-type.p2.dual-d2.text.n-apps\endcsname{15}
\expandafter\def\csname cnt@rev.desc.d7.uiv.tau-T-by-type.p2.dual-d2.text.n-apps\endcsname{15}
\expandafter\def\csname pct@rev.desc.d7.uiv.tau-T-by-type.p2.dual-d2.text.n-apps\endcsname{1500.0}
\expandafter\def\csname ppt@rev.desc.d7.uiv.tau-T-by-type.p2.dual-d2.text.n-apps\endcsname{$+$1500.0}
\expandafter\def\csname val@rev.desc.d7.uiv.tau-T-by-type.p2.dual-d2.text.n-tasks\endcsname{30}
\expandafter\def\csname cnt@rev.desc.d7.uiv.tau-T-by-type.p2.dual-d2.text.n-tasks\endcsname{30}
\expandafter\def\csname pct@rev.desc.d7.uiv.tau-T-by-type.p2.dual-d2.text.n-tasks\endcsname{3000.0}
\expandafter\def\csname ppt@rev.desc.d7.uiv.tau-T-by-type.p2.dual-d2.text.n-tasks\endcsname{$+$3000.0}
\expandafter\def\csname val@rev.desc.d7.uiv.tau-T-by-type.p2.dual-d2.text.tau.ci90.0\endcsname{0.000}
\expandafter\def\csname cnt@rev.desc.d7.uiv.tau-T-by-type.p2.dual-d2.text.tau.ci90.0\endcsname{0}
\expandafter\def\csname pct@rev.desc.d7.uiv.tau-T-by-type.p2.dual-d2.text.tau.ci90.0\endcsname{0.0}
\expandafter\def\csname ppt@rev.desc.d7.uiv.tau-T-by-type.p2.dual-d2.text.tau.ci90.0\endcsname{0.0}
\expandafter\def\csname val@rev.desc.d7.uiv.tau-T-by-type.p2.dual-d2.text.tau.ci90.1\endcsname{0.200}
\expandafter\def\csname cnt@rev.desc.d7.uiv.tau-T-by-type.p2.dual-d2.text.tau.ci90.1\endcsname{0}
\expandafter\def\csname pct@rev.desc.d7.uiv.tau-T-by-type.p2.dual-d2.text.tau.ci90.1\endcsname{20.0}
\expandafter\def\csname ppt@rev.desc.d7.uiv.tau-T-by-type.p2.dual-d2.text.tau.ci90.1\endcsname{$+$20.0}
\expandafter\def\csname val@rev.desc.d7.uiv.tau-T-by-type.p2.dual-d2.text.tau.ci95-excludes-0\endcsname{0}
\expandafter\def\csname cnt@rev.desc.d7.uiv.tau-T-by-type.p2.dual-d2.text.tau.ci95-excludes-0\endcsname{0}
\expandafter\def\csname pct@rev.desc.d7.uiv.tau-T-by-type.p2.dual-d2.text.tau.ci95-excludes-0\endcsname{0.0}
\expandafter\def\csname ppt@rev.desc.d7.uiv.tau-T-by-type.p2.dual-d2.text.tau.ci95-excludes-0\endcsname{0.0}
\expandafter\def\csname val@rev.desc.d7.uiv.tau-T-by-type.p2.dual-d2.text.tau.ci95.0\endcsname{0.000}
\expandafter\def\csname cnt@rev.desc.d7.uiv.tau-T-by-type.p2.dual-d2.text.tau.ci95.0\endcsname{0}
\expandafter\def\csname pct@rev.desc.d7.uiv.tau-T-by-type.p2.dual-d2.text.tau.ci95.0\endcsname{0.0}
\expandafter\def\csname ppt@rev.desc.d7.uiv.tau-T-by-type.p2.dual-d2.text.tau.ci95.0\endcsname{0.0}
\expandafter\def\csname val@rev.desc.d7.uiv.tau-T-by-type.p2.dual-d2.text.tau.ci95.1\endcsname{0.200}
\expandafter\def\csname cnt@rev.desc.d7.uiv.tau-T-by-type.p2.dual-d2.text.tau.ci95.1\endcsname{0}
\expandafter\def\csname pct@rev.desc.d7.uiv.tau-T-by-type.p2.dual-d2.text.tau.ci95.1\endcsname{20.0}
\expandafter\def\csname ppt@rev.desc.d7.uiv.tau-T-by-type.p2.dual-d2.text.tau.ci95.1\endcsname{$+$20.0}
\expandafter\def\csname val@rev.desc.d7.uiv.tau-T-by-type.p2.dual-d2.text.tau.mean\endcsname{0.067}
\expandafter\def\csname cnt@rev.desc.d7.uiv.tau-T-by-type.p2.dual-d2.text.tau.mean\endcsname{0}
\expandafter\def\csname pct@rev.desc.d7.uiv.tau-T-by-type.p2.dual-d2.text.tau.mean\endcsname{6.7}
\expandafter\def\csname ppt@rev.desc.d7.uiv.tau-T-by-type.p2.dual-d2.text.tau.mean\endcsname{$+$6.7}
\expandafter\def\csname val@rev.desc.d7.uiv.tau-T-by-type.p2.dual-d2.text.tau.n-apps\endcsname{15}
\expandafter\def\csname cnt@rev.desc.d7.uiv.tau-T-by-type.p2.dual-d2.text.tau.n-apps\endcsname{15}
\expandafter\def\csname pct@rev.desc.d7.uiv.tau-T-by-type.p2.dual-d2.text.tau.n-apps\endcsname{1500.0}
\expandafter\def\csname ppt@rev.desc.d7.uiv.tau-T-by-type.p2.dual-d2.text.tau.n-apps\endcsname{$+$1500.0}
\expandafter\def\csname val@rev.desc.d7.uiv.tau-T-by-type.p2.dual-d2.text.tau.tost1.ci90-inside\endcsname{0}
\expandafter\def\csname cnt@rev.desc.d7.uiv.tau-T-by-type.p2.dual-d2.text.tau.tost1.ci90-inside\endcsname{0}
\expandafter\def\csname pct@rev.desc.d7.uiv.tau-T-by-type.p2.dual-d2.text.tau.tost1.ci90-inside\endcsname{0.0}
\expandafter\def\csname ppt@rev.desc.d7.uiv.tau-T-by-type.p2.dual-d2.text.tau.tost1.ci90-inside\endcsname{0.0}
\expandafter\def\csname val@rev.desc.d7.uiv.tau-T-by-type.p2.dual-d2.text.tau.tost1.equivalent\endcsname{0}
\expandafter\def\csname cnt@rev.desc.d7.uiv.tau-T-by-type.p2.dual-d2.text.tau.tost1.equivalent\endcsname{0}
\expandafter\def\csname pct@rev.desc.d7.uiv.tau-T-by-type.p2.dual-d2.text.tau.tost1.equivalent\endcsname{0.0}
\expandafter\def\csname ppt@rev.desc.d7.uiv.tau-T-by-type.p2.dual-d2.text.tau.tost1.equivalent\endcsname{0.0}
\expandafter\def\csname val@rev.desc.d7.uiv.tau-T-by-type.p2.dual-d2.text.tau.tost1.p\endcsname{0.647}
\expandafter\def\csname cnt@rev.desc.d7.uiv.tau-T-by-type.p2.dual-d2.text.tau.tost1.p\endcsname{1}
\expandafter\def\csname pct@rev.desc.d7.uiv.tau-T-by-type.p2.dual-d2.text.tau.tost1.p\endcsname{64.7}
\expandafter\def\csname ppt@rev.desc.d7.uiv.tau-T-by-type.p2.dual-d2.text.tau.tost1.p\endcsname{$+$64.7}
\expandafter\def\csname val@rev.desc.d7.uiv.tau-T-by-type.p2.dual-d2.text.tau.tost5.ci90-inside\endcsname{0}
\expandafter\def\csname cnt@rev.desc.d7.uiv.tau-T-by-type.p2.dual-d2.text.tau.tost5.ci90-inside\endcsname{0}
\expandafter\def\csname pct@rev.desc.d7.uiv.tau-T-by-type.p2.dual-d2.text.tau.tost5.ci90-inside\endcsname{0.0}
\expandafter\def\csname ppt@rev.desc.d7.uiv.tau-T-by-type.p2.dual-d2.text.tau.tost5.ci90-inside\endcsname{0.0}
\expandafter\def\csname val@rev.desc.d7.uiv.tau-T-by-type.p2.dual-d2.text.tau.tost5.equivalent\endcsname{0}
\expandafter\def\csname cnt@rev.desc.d7.uiv.tau-T-by-type.p2.dual-d2.text.tau.tost5.equivalent\endcsname{0}
\expandafter\def\csname pct@rev.desc.d7.uiv.tau-T-by-type.p2.dual-d2.text.tau.tost5.equivalent\endcsname{0.0}
\expandafter\def\csname ppt@rev.desc.d7.uiv.tau-T-by-type.p2.dual-d2.text.tau.tost5.equivalent\endcsname{0.0}
\expandafter\def\csname val@rev.desc.d7.uiv.tau-T-by-type.p2.dual-d2.text.tau.tost5.p\endcsname{0.647}
\expandafter\def\csname cnt@rev.desc.d7.uiv.tau-T-by-type.p2.dual-d2.text.tau.tost5.p\endcsname{1}
\expandafter\def\csname pct@rev.desc.d7.uiv.tau-T-by-type.p2.dual-d2.text.tau.tost5.p\endcsname{64.7}
\expandafter\def\csname ppt@rev.desc.d7.uiv.tau-T-by-type.p2.dual-d2.text.tau.tost5.p\endcsname{$+$64.7}
\expandafter\def\csname val@rev.desc.d7.uiv.tau-T-by-type.p2.grid2-d3.icon-minus-text-on-apps-with-both.ci90.0\endcsname{-0.121}
\expandafter\def\csname cnt@rev.desc.d7.uiv.tau-T-by-type.p2.grid2-d3.icon-minus-text-on-apps-with-both.ci90.0\endcsname{0}
\expandafter\def\csname pct@rev.desc.d7.uiv.tau-T-by-type.p2.grid2-d3.icon-minus-text-on-apps-with-both.ci90.0\endcsname{$-$12.1}
\expandafter\def\csname ppt@rev.desc.d7.uiv.tau-T-by-type.p2.grid2-d3.icon-minus-text-on-apps-with-both.ci90.0\endcsname{$-$12.1}
\expandafter\def\csname val@rev.desc.d7.uiv.tau-T-by-type.p2.grid2-d3.icon-minus-text-on-apps-with-both.ci90.1\endcsname{0.014}
\expandafter\def\csname cnt@rev.desc.d7.uiv.tau-T-by-type.p2.grid2-d3.icon-minus-text-on-apps-with-both.ci90.1\endcsname{0}
\expandafter\def\csname pct@rev.desc.d7.uiv.tau-T-by-type.p2.grid2-d3.icon-minus-text-on-apps-with-both.ci90.1\endcsname{1.4}
\expandafter\def\csname ppt@rev.desc.d7.uiv.tau-T-by-type.p2.grid2-d3.icon-minus-text-on-apps-with-both.ci90.1\endcsname{$+$1.4}
\expandafter\def\csname val@rev.desc.d7.uiv.tau-T-by-type.p2.grid2-d3.icon-minus-text-on-apps-with-both.ci95-excludes-0\endcsname{0}
\expandafter\def\csname cnt@rev.desc.d7.uiv.tau-T-by-type.p2.grid2-d3.icon-minus-text-on-apps-with-both.ci95-excludes-0\endcsname{0}
\expandafter\def\csname pct@rev.desc.d7.uiv.tau-T-by-type.p2.grid2-d3.icon-minus-text-on-apps-with-both.ci95-excludes-0\endcsname{0.0}
\expandafter\def\csname ppt@rev.desc.d7.uiv.tau-T-by-type.p2.grid2-d3.icon-minus-text-on-apps-with-both.ci95-excludes-0\endcsname{0.0}
\expandafter\def\csname val@rev.desc.d7.uiv.tau-T-by-type.p2.grid2-d3.icon-minus-text-on-apps-with-both.ci95.0\endcsname{-0.143}
\expandafter\def\csname cnt@rev.desc.d7.uiv.tau-T-by-type.p2.grid2-d3.icon-minus-text-on-apps-with-both.ci95.0\endcsname{0}
\expandafter\def\csname pct@rev.desc.d7.uiv.tau-T-by-type.p2.grid2-d3.icon-minus-text-on-apps-with-both.ci95.0\endcsname{$-$14.3}
\expandafter\def\csname ppt@rev.desc.d7.uiv.tau-T-by-type.p2.grid2-d3.icon-minus-text-on-apps-with-both.ci95.0\endcsname{$-$14.3}
\expandafter\def\csname val@rev.desc.d7.uiv.tau-T-by-type.p2.grid2-d3.icon-minus-text-on-apps-with-both.ci95.1\endcsname{0.021}
\expandafter\def\csname cnt@rev.desc.d7.uiv.tau-T-by-type.p2.grid2-d3.icon-minus-text-on-apps-with-both.ci95.1\endcsname{0}
\expandafter\def\csname pct@rev.desc.d7.uiv.tau-T-by-type.p2.grid2-d3.icon-minus-text-on-apps-with-both.ci95.1\endcsname{2.1}
\expandafter\def\csname ppt@rev.desc.d7.uiv.tau-T-by-type.p2.grid2-d3.icon-minus-text-on-apps-with-both.ci95.1\endcsname{$+$2.1}
\expandafter\def\csname val@rev.desc.d7.uiv.tau-T-by-type.p2.grid2-d3.icon-minus-text-on-apps-with-both.mean\endcsname{-0.050}
\expandafter\def\csname cnt@rev.desc.d7.uiv.tau-T-by-type.p2.grid2-d3.icon-minus-text-on-apps-with-both.mean\endcsname{0}
\expandafter\def\csname pct@rev.desc.d7.uiv.tau-T-by-type.p2.grid2-d3.icon-minus-text-on-apps-with-both.mean\endcsname{$-$5.0}
\expandafter\def\csname ppt@rev.desc.d7.uiv.tau-T-by-type.p2.grid2-d3.icon-minus-text-on-apps-with-both.mean\endcsname{$-$5.0}
\expandafter\def\csname val@rev.desc.d7.uiv.tau-T-by-type.p2.grid2-d3.icon-minus-text-on-apps-with-both.n-apps\endcsname{14}
\expandafter\def\csname cnt@rev.desc.d7.uiv.tau-T-by-type.p2.grid2-d3.icon-minus-text-on-apps-with-both.n-apps\endcsname{14}
\expandafter\def\csname pct@rev.desc.d7.uiv.tau-T-by-type.p2.grid2-d3.icon-minus-text-on-apps-with-both.n-apps\endcsname{1400.0}
\expandafter\def\csname ppt@rev.desc.d7.uiv.tau-T-by-type.p2.grid2-d3.icon-minus-text-on-apps-with-both.n-apps\endcsname{$+$1400.0}
\expandafter\def\csname val@rev.desc.d7.uiv.tau-T-by-type.p2.grid2-d3.icon-minus-text-on-apps-with-both.tost1.ci90-inside\endcsname{0}
\expandafter\def\csname cnt@rev.desc.d7.uiv.tau-T-by-type.p2.grid2-d3.icon-minus-text-on-apps-with-both.tost1.ci90-inside\endcsname{0}
\expandafter\def\csname pct@rev.desc.d7.uiv.tau-T-by-type.p2.grid2-d3.icon-minus-text-on-apps-with-both.tost1.ci90-inside\endcsname{0.0}
\expandafter\def\csname ppt@rev.desc.d7.uiv.tau-T-by-type.p2.grid2-d3.icon-minus-text-on-apps-with-both.tost1.ci90-inside\endcsname{0.0}
\expandafter\def\csname val@rev.desc.d7.uiv.tau-T-by-type.p2.grid2-d3.icon-minus-text-on-apps-with-both.tost1.equivalent\endcsname{0}
\expandafter\def\csname cnt@rev.desc.d7.uiv.tau-T-by-type.p2.grid2-d3.icon-minus-text-on-apps-with-both.tost1.equivalent\endcsname{0}
\expandafter\def\csname pct@rev.desc.d7.uiv.tau-T-by-type.p2.grid2-d3.icon-minus-text-on-apps-with-both.tost1.equivalent\endcsname{0.0}
\expandafter\def\csname ppt@rev.desc.d7.uiv.tau-T-by-type.p2.grid2-d3.icon-minus-text-on-apps-with-both.tost1.equivalent\endcsname{0.0}
\expandafter\def\csname val@rev.desc.d7.uiv.tau-T-by-type.p2.grid2-d3.icon-minus-text-on-apps-with-both.tost1.p\endcsname{0.841}
\expandafter\def\csname cnt@rev.desc.d7.uiv.tau-T-by-type.p2.grid2-d3.icon-minus-text-on-apps-with-both.tost1.p\endcsname{1}
\expandafter\def\csname pct@rev.desc.d7.uiv.tau-T-by-type.p2.grid2-d3.icon-minus-text-on-apps-with-both.tost1.p\endcsname{84.1}
\expandafter\def\csname ppt@rev.desc.d7.uiv.tau-T-by-type.p2.grid2-d3.icon-minus-text-on-apps-with-both.tost1.p\endcsname{$+$84.1}
\expandafter\def\csname val@rev.desc.d7.uiv.tau-T-by-type.p2.grid2-d3.icon-minus-text-on-apps-with-both.tost5.ci90-inside\endcsname{0}
\expandafter\def\csname cnt@rev.desc.d7.uiv.tau-T-by-type.p2.grid2-d3.icon-minus-text-on-apps-with-both.tost5.ci90-inside\endcsname{0}
\expandafter\def\csname pct@rev.desc.d7.uiv.tau-T-by-type.p2.grid2-d3.icon-minus-text-on-apps-with-both.tost5.ci90-inside\endcsname{0.0}
\expandafter\def\csname ppt@rev.desc.d7.uiv.tau-T-by-type.p2.grid2-d3.icon-minus-text-on-apps-with-both.tost5.ci90-inside\endcsname{0.0}
\expandafter\def\csname val@rev.desc.d7.uiv.tau-T-by-type.p2.grid2-d3.icon-minus-text-on-apps-with-both.tost5.equivalent\endcsname{0}
\expandafter\def\csname cnt@rev.desc.d7.uiv.tau-T-by-type.p2.grid2-d3.icon-minus-text-on-apps-with-both.tost5.equivalent\endcsname{0}
\expandafter\def\csname pct@rev.desc.d7.uiv.tau-T-by-type.p2.grid2-d3.icon-minus-text-on-apps-with-both.tost5.equivalent\endcsname{0.0}
\expandafter\def\csname ppt@rev.desc.d7.uiv.tau-T-by-type.p2.grid2-d3.icon-minus-text-on-apps-with-both.tost5.equivalent\endcsname{0.0}
\expandafter\def\csname val@rev.desc.d7.uiv.tau-T-by-type.p2.grid2-d3.icon-minus-text-on-apps-with-both.tost5.p\endcsname{0.429}
\expandafter\def\csname cnt@rev.desc.d7.uiv.tau-T-by-type.p2.grid2-d3.icon-minus-text-on-apps-with-both.tost5.p\endcsname{0}
\expandafter\def\csname pct@rev.desc.d7.uiv.tau-T-by-type.p2.grid2-d3.icon-minus-text-on-apps-with-both.tost5.p\endcsname{42.9}
\expandafter\def\csname ppt@rev.desc.d7.uiv.tau-T-by-type.p2.grid2-d3.icon-minus-text-on-apps-with-both.tost5.p\endcsname{$+$42.9}
\expandafter\def\csname val@rev.desc.d7.uiv.tau-T-by-type.p2.grid2-d3.icon.acc-F\endcsname{0.833}
\expandafter\def\csname cnt@rev.desc.d7.uiv.tau-T-by-type.p2.grid2-d3.icon.acc-F\endcsname{1}
\expandafter\def\csname pct@rev.desc.d7.uiv.tau-T-by-type.p2.grid2-d3.icon.acc-F\endcsname{83.3}
\expandafter\def\csname ppt@rev.desc.d7.uiv.tau-T-by-type.p2.grid2-d3.icon.acc-F\endcsname{$+$83.3}
\expandafter\def\csname val@rev.desc.d7.uiv.tau-T-by-type.p2.grid2-d3.icon.acc-direct\endcsname{0.851}
\expandafter\def\csname cnt@rev.desc.d7.uiv.tau-T-by-type.p2.grid2-d3.icon.acc-direct\endcsname{1}
\expandafter\def\csname pct@rev.desc.d7.uiv.tau-T-by-type.p2.grid2-d3.icon.acc-direct\endcsname{85.1}
\expandafter\def\csname ppt@rev.desc.d7.uiv.tau-T-by-type.p2.grid2-d3.icon.acc-direct\endcsname{$+$85.1}
\expandafter\def\csname val@rev.desc.d7.uiv.tau-T-by-type.p2.grid2-d3.icon.n-apps\endcsname{23}
\expandafter\def\csname cnt@rev.desc.d7.uiv.tau-T-by-type.p2.grid2-d3.icon.n-apps\endcsname{23}
\expandafter\def\csname pct@rev.desc.d7.uiv.tau-T-by-type.p2.grid2-d3.icon.n-apps\endcsname{2300.0}
\expandafter\def\csname ppt@rev.desc.d7.uiv.tau-T-by-type.p2.grid2-d3.icon.n-apps\endcsname{$+$2300.0}
\expandafter\def\csname val@rev.desc.d7.uiv.tau-T-by-type.p2.grid2-d3.icon.n-tasks\endcsname{114}
\expandafter\def\csname cnt@rev.desc.d7.uiv.tau-T-by-type.p2.grid2-d3.icon.n-tasks\endcsname{114}
\expandafter\def\csname pct@rev.desc.d7.uiv.tau-T-by-type.p2.grid2-d3.icon.n-tasks\endcsname{11400.0}
\expandafter\def\csname ppt@rev.desc.d7.uiv.tau-T-by-type.p2.grid2-d3.icon.n-tasks\endcsname{$+$11400.0}
\expandafter\def\csname val@rev.desc.d7.uiv.tau-T-by-type.p2.grid2-d3.icon.tau.ci90.0\endcsname{-0.048}
\expandafter\def\csname cnt@rev.desc.d7.uiv.tau-T-by-type.p2.grid2-d3.icon.tau.ci90.0\endcsname{0}
\expandafter\def\csname pct@rev.desc.d7.uiv.tau-T-by-type.p2.grid2-d3.icon.tau.ci90.0\endcsname{$-$4.8}
\expandafter\def\csname ppt@rev.desc.d7.uiv.tau-T-by-type.p2.grid2-d3.icon.tau.ci90.0\endcsname{$-$4.8}
\expandafter\def\csname val@rev.desc.d7.uiv.tau-T-by-type.p2.grid2-d3.icon.tau.ci90.1\endcsname{0.016}
\expandafter\def\csname cnt@rev.desc.d7.uiv.tau-T-by-type.p2.grid2-d3.icon.tau.ci90.1\endcsname{0}
\expandafter\def\csname pct@rev.desc.d7.uiv.tau-T-by-type.p2.grid2-d3.icon.tau.ci90.1\endcsname{1.6}
\expandafter\def\csname ppt@rev.desc.d7.uiv.tau-T-by-type.p2.grid2-d3.icon.tau.ci90.1\endcsname{$+$1.6}
\expandafter\def\csname val@rev.desc.d7.uiv.tau-T-by-type.p2.grid2-d3.icon.tau.ci95-excludes-0\endcsname{0}
\expandafter\def\csname cnt@rev.desc.d7.uiv.tau-T-by-type.p2.grid2-d3.icon.tau.ci95-excludes-0\endcsname{0}
\expandafter\def\csname pct@rev.desc.d7.uiv.tau-T-by-type.p2.grid2-d3.icon.tau.ci95-excludes-0\endcsname{0.0}
\expandafter\def\csname ppt@rev.desc.d7.uiv.tau-T-by-type.p2.grid2-d3.icon.tau.ci95-excludes-0\endcsname{0.0}
\expandafter\def\csname val@rev.desc.d7.uiv.tau-T-by-type.p2.grid2-d3.icon.tau.ci95.0\endcsname{-0.055}
\expandafter\def\csname cnt@rev.desc.d7.uiv.tau-T-by-type.p2.grid2-d3.icon.tau.ci95.0\endcsname{0}
\expandafter\def\csname pct@rev.desc.d7.uiv.tau-T-by-type.p2.grid2-d3.icon.tau.ci95.0\endcsname{$-$5.5}
\expandafter\def\csname ppt@rev.desc.d7.uiv.tau-T-by-type.p2.grid2-d3.icon.tau.ci95.0\endcsname{$-$5.5}
\expandafter\def\csname val@rev.desc.d7.uiv.tau-T-by-type.p2.grid2-d3.icon.tau.ci95.1\endcsname{0.022}
\expandafter\def\csname cnt@rev.desc.d7.uiv.tau-T-by-type.p2.grid2-d3.icon.tau.ci95.1\endcsname{0}
\expandafter\def\csname pct@rev.desc.d7.uiv.tau-T-by-type.p2.grid2-d3.icon.tau.ci95.1\endcsname{2.2}
\expandafter\def\csname ppt@rev.desc.d7.uiv.tau-T-by-type.p2.grid2-d3.icon.tau.ci95.1\endcsname{$+$2.2}
\expandafter\def\csname val@rev.desc.d7.uiv.tau-T-by-type.p2.grid2-d3.icon.tau.mean\endcsname{-0.016}
\expandafter\def\csname cnt@rev.desc.d7.uiv.tau-T-by-type.p2.grid2-d3.icon.tau.mean\endcsname{0}
\expandafter\def\csname pct@rev.desc.d7.uiv.tau-T-by-type.p2.grid2-d3.icon.tau.mean\endcsname{$-$1.6}
\expandafter\def\csname ppt@rev.desc.d7.uiv.tau-T-by-type.p2.grid2-d3.icon.tau.mean\endcsname{$-$1.6}
\expandafter\def\csname val@rev.desc.d7.uiv.tau-T-by-type.p2.grid2-d3.icon.tau.n-apps\endcsname{23}
\expandafter\def\csname cnt@rev.desc.d7.uiv.tau-T-by-type.p2.grid2-d3.icon.tau.n-apps\endcsname{23}
\expandafter\def\csname pct@rev.desc.d7.uiv.tau-T-by-type.p2.grid2-d3.icon.tau.n-apps\endcsname{2300.0}
\expandafter\def\csname ppt@rev.desc.d7.uiv.tau-T-by-type.p2.grid2-d3.icon.tau.n-apps\endcsname{$+$2300.0}
\expandafter\def\csname val@rev.desc.d7.uiv.tau-T-by-type.p2.grid2-d3.icon.tau.tost1.ci90-inside\endcsname{0}
\expandafter\def\csname cnt@rev.desc.d7.uiv.tau-T-by-type.p2.grid2-d3.icon.tau.tost1.ci90-inside\endcsname{0}
\expandafter\def\csname pct@rev.desc.d7.uiv.tau-T-by-type.p2.grid2-d3.icon.tau.tost1.ci90-inside\endcsname{0.0}
\expandafter\def\csname ppt@rev.desc.d7.uiv.tau-T-by-type.p2.grid2-d3.icon.tau.tost1.ci90-inside\endcsname{0.0}
\expandafter\def\csname val@rev.desc.d7.uiv.tau-T-by-type.p2.grid2-d3.icon.tau.tost1.equivalent\endcsname{0}
\expandafter\def\csname cnt@rev.desc.d7.uiv.tau-T-by-type.p2.grid2-d3.icon.tau.tost1.equivalent\endcsname{0}
\expandafter\def\csname pct@rev.desc.d7.uiv.tau-T-by-type.p2.grid2-d3.icon.tau.tost1.equivalent\endcsname{0.0}
\expandafter\def\csname ppt@rev.desc.d7.uiv.tau-T-by-type.p2.grid2-d3.icon.tau.tost1.equivalent\endcsname{0.0}
\expandafter\def\csname val@rev.desc.d7.uiv.tau-T-by-type.p2.grid2-d3.icon.tau.tost1.p\endcsname{0.621}
\expandafter\def\csname cnt@rev.desc.d7.uiv.tau-T-by-type.p2.grid2-d3.icon.tau.tost1.p\endcsname{1}
\expandafter\def\csname pct@rev.desc.d7.uiv.tau-T-by-type.p2.grid2-d3.icon.tau.tost1.p\endcsname{62.1}
\expandafter\def\csname ppt@rev.desc.d7.uiv.tau-T-by-type.p2.grid2-d3.icon.tau.tost1.p\endcsname{$+$62.1}
\expandafter\def\csname val@rev.desc.d7.uiv.tau-T-by-type.p2.grid2-d3.icon.tau.tost5.ci90-inside\endcsname{1}
\expandafter\def\csname cnt@rev.desc.d7.uiv.tau-T-by-type.p2.grid2-d3.icon.tau.tost5.ci90-inside\endcsname{1}
\expandafter\def\csname pct@rev.desc.d7.uiv.tau-T-by-type.p2.grid2-d3.icon.tau.tost5.ci90-inside\endcsname{100.0}
\expandafter\def\csname ppt@rev.desc.d7.uiv.tau-T-by-type.p2.grid2-d3.icon.tau.tost5.ci90-inside\endcsname{$+$100.0}
\expandafter\def\csname val@rev.desc.d7.uiv.tau-T-by-type.p2.grid2-d3.icon.tau.tost5.equivalent\endcsname{1}
\expandafter\def\csname cnt@rev.desc.d7.uiv.tau-T-by-type.p2.grid2-d3.icon.tau.tost5.equivalent\endcsname{1}
\expandafter\def\csname pct@rev.desc.d7.uiv.tau-T-by-type.p2.grid2-d3.icon.tau.tost5.equivalent\endcsname{100.0}
\expandafter\def\csname ppt@rev.desc.d7.uiv.tau-T-by-type.p2.grid2-d3.icon.tau.tost5.equivalent\endcsname{$+$100.0}
\expandafter\def\csname val@rev.desc.d7.uiv.tau-T-by-type.p2.grid2-d3.icon.tau.tost5.p\endcsname{0.039}
\expandafter\def\csname cnt@rev.desc.d7.uiv.tau-T-by-type.p2.grid2-d3.icon.tau.tost5.p\endcsname{0}
\expandafter\def\csname pct@rev.desc.d7.uiv.tau-T-by-type.p2.grid2-d3.icon.tau.tost5.p\endcsname{3.9}
\expandafter\def\csname ppt@rev.desc.d7.uiv.tau-T-by-type.p2.grid2-d3.icon.tau.tost5.p\endcsname{$+$3.9}
\expandafter\def\csname val@rev.desc.d7.uiv.tau-T-by-type.p2.grid2-d3.text.acc-F\endcsname{0.833}
\expandafter\def\csname cnt@rev.desc.d7.uiv.tau-T-by-type.p2.grid2-d3.text.acc-F\endcsname{1}
\expandafter\def\csname pct@rev.desc.d7.uiv.tau-T-by-type.p2.grid2-d3.text.acc-F\endcsname{83.3}
\expandafter\def\csname ppt@rev.desc.d7.uiv.tau-T-by-type.p2.grid2-d3.text.acc-F\endcsname{$+$83.3}
\expandafter\def\csname val@rev.desc.d7.uiv.tau-T-by-type.p2.grid2-d3.text.acc-direct\endcsname{0.800}
\expandafter\def\csname cnt@rev.desc.d7.uiv.tau-T-by-type.p2.grid2-d3.text.acc-direct\endcsname{1}
\expandafter\def\csname pct@rev.desc.d7.uiv.tau-T-by-type.p2.grid2-d3.text.acc-direct\endcsname{80.0}
\expandafter\def\csname ppt@rev.desc.d7.uiv.tau-T-by-type.p2.grid2-d3.text.acc-direct\endcsname{$+$80.0}
\expandafter\def\csname val@rev.desc.d7.uiv.tau-T-by-type.p2.grid2-d3.text.n-apps\endcsname{15}
\expandafter\def\csname cnt@rev.desc.d7.uiv.tau-T-by-type.p2.grid2-d3.text.n-apps\endcsname{15}
\expandafter\def\csname pct@rev.desc.d7.uiv.tau-T-by-type.p2.grid2-d3.text.n-apps\endcsname{1500.0}
\expandafter\def\csname ppt@rev.desc.d7.uiv.tau-T-by-type.p2.grid2-d3.text.n-apps\endcsname{$+$1500.0}
\expandafter\def\csname val@rev.desc.d7.uiv.tau-T-by-type.p2.grid2-d3.text.n-tasks\endcsname{30}
\expandafter\def\csname cnt@rev.desc.d7.uiv.tau-T-by-type.p2.grid2-d3.text.n-tasks\endcsname{30}
\expandafter\def\csname pct@rev.desc.d7.uiv.tau-T-by-type.p2.grid2-d3.text.n-tasks\endcsname{3000.0}
\expandafter\def\csname ppt@rev.desc.d7.uiv.tau-T-by-type.p2.grid2-d3.text.n-tasks\endcsname{$+$3000.0}
\expandafter\def\csname val@rev.desc.d7.uiv.tau-T-by-type.p2.grid2-d3.text.tau.ci90.0\endcsname{0.000}
\expandafter\def\csname cnt@rev.desc.d7.uiv.tau-T-by-type.p2.grid2-d3.text.tau.ci90.0\endcsname{0}
\expandafter\def\csname pct@rev.desc.d7.uiv.tau-T-by-type.p2.grid2-d3.text.tau.ci90.0\endcsname{0.0}
\expandafter\def\csname ppt@rev.desc.d7.uiv.tau-T-by-type.p2.grid2-d3.text.tau.ci90.0\endcsname{0.0}
\expandafter\def\csname val@rev.desc.d7.uiv.tau-T-by-type.p2.grid2-d3.text.tau.ci90.1\endcsname{0.100}
\expandafter\def\csname cnt@rev.desc.d7.uiv.tau-T-by-type.p2.grid2-d3.text.tau.ci90.1\endcsname{0}
\expandafter\def\csname pct@rev.desc.d7.uiv.tau-T-by-type.p2.grid2-d3.text.tau.ci90.1\endcsname{10.0}
\expandafter\def\csname ppt@rev.desc.d7.uiv.tau-T-by-type.p2.grid2-d3.text.tau.ci90.1\endcsname{$+$10.0}
\expandafter\def\csname val@rev.desc.d7.uiv.tau-T-by-type.p2.grid2-d3.text.tau.ci95-excludes-0\endcsname{0}
\expandafter\def\csname cnt@rev.desc.d7.uiv.tau-T-by-type.p2.grid2-d3.text.tau.ci95-excludes-0\endcsname{0}
\expandafter\def\csname pct@rev.desc.d7.uiv.tau-T-by-type.p2.grid2-d3.text.tau.ci95-excludes-0\endcsname{0.0}
\expandafter\def\csname ppt@rev.desc.d7.uiv.tau-T-by-type.p2.grid2-d3.text.tau.ci95-excludes-0\endcsname{0.0}
\expandafter\def\csname val@rev.desc.d7.uiv.tau-T-by-type.p2.grid2-d3.text.tau.ci95.0\endcsname{0.000}
\expandafter\def\csname cnt@rev.desc.d7.uiv.tau-T-by-type.p2.grid2-d3.text.tau.ci95.0\endcsname{0}
\expandafter\def\csname pct@rev.desc.d7.uiv.tau-T-by-type.p2.grid2-d3.text.tau.ci95.0\endcsname{0.0}
\expandafter\def\csname ppt@rev.desc.d7.uiv.tau-T-by-type.p2.grid2-d3.text.tau.ci95.0\endcsname{0.0}
\expandafter\def\csname val@rev.desc.d7.uiv.tau-T-by-type.p2.grid2-d3.text.tau.ci95.1\endcsname{0.100}
\expandafter\def\csname cnt@rev.desc.d7.uiv.tau-T-by-type.p2.grid2-d3.text.tau.ci95.1\endcsname{0}
\expandafter\def\csname pct@rev.desc.d7.uiv.tau-T-by-type.p2.grid2-d3.text.tau.ci95.1\endcsname{10.0}
\expandafter\def\csname ppt@rev.desc.d7.uiv.tau-T-by-type.p2.grid2-d3.text.tau.ci95.1\endcsname{$+$10.0}
\expandafter\def\csname val@rev.desc.d7.uiv.tau-T-by-type.p2.grid2-d3.text.tau.mean\endcsname{0.033}
\expandafter\def\csname cnt@rev.desc.d7.uiv.tau-T-by-type.p2.grid2-d3.text.tau.mean\endcsname{0}
\expandafter\def\csname pct@rev.desc.d7.uiv.tau-T-by-type.p2.grid2-d3.text.tau.mean\endcsname{3.3}
\expandafter\def\csname ppt@rev.desc.d7.uiv.tau-T-by-type.p2.grid2-d3.text.tau.mean\endcsname{$+$3.3}
\expandafter\def\csname val@rev.desc.d7.uiv.tau-T-by-type.p2.grid2-d3.text.tau.n-apps\endcsname{15}
\expandafter\def\csname cnt@rev.desc.d7.uiv.tau-T-by-type.p2.grid2-d3.text.tau.n-apps\endcsname{15}
\expandafter\def\csname pct@rev.desc.d7.uiv.tau-T-by-type.p2.grid2-d3.text.tau.n-apps\endcsname{1500.0}
\expandafter\def\csname ppt@rev.desc.d7.uiv.tau-T-by-type.p2.grid2-d3.text.tau.n-apps\endcsname{$+$1500.0}
\expandafter\def\csname val@rev.desc.d7.uiv.tau-T-by-type.p2.grid2-d3.text.tau.tost1.ci90-inside\endcsname{0}
\expandafter\def\csname cnt@rev.desc.d7.uiv.tau-T-by-type.p2.grid2-d3.text.tau.tost1.ci90-inside\endcsname{0}
\expandafter\def\csname pct@rev.desc.d7.uiv.tau-T-by-type.p2.grid2-d3.text.tau.tost1.ci90-inside\endcsname{0.0}
\expandafter\def\csname ppt@rev.desc.d7.uiv.tau-T-by-type.p2.grid2-d3.text.tau.tost1.ci90-inside\endcsname{0.0}
\expandafter\def\csname val@rev.desc.d7.uiv.tau-T-by-type.p2.grid2-d3.text.tau.tost1.equivalent\endcsname{0}
\expandafter\def\csname cnt@rev.desc.d7.uiv.tau-T-by-type.p2.grid2-d3.text.tau.tost1.equivalent\endcsname{0}
\expandafter\def\csname pct@rev.desc.d7.uiv.tau-T-by-type.p2.grid2-d3.text.tau.tost1.equivalent\endcsname{0.0}
\expandafter\def\csname ppt@rev.desc.d7.uiv.tau-T-by-type.p2.grid2-d3.text.tau.tost1.equivalent\endcsname{0.0}
\expandafter\def\csname val@rev.desc.d7.uiv.tau-T-by-type.p2.grid2-d3.text.tau.tost1.p\endcsname{0.656}
\expandafter\def\csname cnt@rev.desc.d7.uiv.tau-T-by-type.p2.grid2-d3.text.tau.tost1.p\endcsname{1}
\expandafter\def\csname pct@rev.desc.d7.uiv.tau-T-by-type.p2.grid2-d3.text.tau.tost1.p\endcsname{65.6}
\expandafter\def\csname ppt@rev.desc.d7.uiv.tau-T-by-type.p2.grid2-d3.text.tau.tost1.p\endcsname{$+$65.6}
\expandafter\def\csname val@rev.desc.d7.uiv.tau-T-by-type.p2.grid2-d3.text.tau.tost5.ci90-inside\endcsname{0}
\expandafter\def\csname cnt@rev.desc.d7.uiv.tau-T-by-type.p2.grid2-d3.text.tau.tost5.ci90-inside\endcsname{0}
\expandafter\def\csname pct@rev.desc.d7.uiv.tau-T-by-type.p2.grid2-d3.text.tau.tost5.ci90-inside\endcsname{0.0}
\expandafter\def\csname ppt@rev.desc.d7.uiv.tau-T-by-type.p2.grid2-d3.text.tau.tost5.ci90-inside\endcsname{0.0}
\expandafter\def\csname val@rev.desc.d7.uiv.tau-T-by-type.p2.grid2-d3.text.tau.tost5.equivalent\endcsname{0}
\expandafter\def\csname cnt@rev.desc.d7.uiv.tau-T-by-type.p2.grid2-d3.text.tau.tost5.equivalent\endcsname{0}
\expandafter\def\csname pct@rev.desc.d7.uiv.tau-T-by-type.p2.grid2-d3.text.tau.tost5.equivalent\endcsname{0.0}
\expandafter\def\csname ppt@rev.desc.d7.uiv.tau-T-by-type.p2.grid2-d3.text.tau.tost5.equivalent\endcsname{0.0}
\expandafter\def\csname val@rev.desc.d7.uiv.tau-T-by-type.p2.grid2-d3.text.tau.tost5.p\endcsname{0.267}
\expandafter\def\csname cnt@rev.desc.d7.uiv.tau-T-by-type.p2.grid2-d3.text.tau.tost5.p\endcsname{0}
\expandafter\def\csname pct@rev.desc.d7.uiv.tau-T-by-type.p2.grid2-d3.text.tau.tost5.p\endcsname{26.7}
\expandafter\def\csname ppt@rev.desc.d7.uiv.tau-T-by-type.p2.grid2-d3.text.tau.tost5.p\endcsname{$+$26.7}
\expandafter\def\csname val@rev.desc.d7.uiv.tau-T-by-type.p2.zoomclick-d3.icon-minus-text-on-apps-with-both.ci90.0\endcsname{-0.143}
\expandafter\def\csname cnt@rev.desc.d7.uiv.tau-T-by-type.p2.zoomclick-d3.icon-minus-text-on-apps-with-both.ci90.0\endcsname{0}
\expandafter\def\csname pct@rev.desc.d7.uiv.tau-T-by-type.p2.zoomclick-d3.icon-minus-text-on-apps-with-both.ci90.0\endcsname{$-$14.3}
\expandafter\def\csname ppt@rev.desc.d7.uiv.tau-T-by-type.p2.zoomclick-d3.icon-minus-text-on-apps-with-both.ci90.0\endcsname{$-$14.3}
\expandafter\def\csname val@rev.desc.d7.uiv.tau-T-by-type.p2.zoomclick-d3.icon-minus-text-on-apps-with-both.ci90.1\endcsname{0.171}
\expandafter\def\csname cnt@rev.desc.d7.uiv.tau-T-by-type.p2.zoomclick-d3.icon-minus-text-on-apps-with-both.ci90.1\endcsname{0}
\expandafter\def\csname pct@rev.desc.d7.uiv.tau-T-by-type.p2.zoomclick-d3.icon-minus-text-on-apps-with-both.ci90.1\endcsname{17.1}
\expandafter\def\csname ppt@rev.desc.d7.uiv.tau-T-by-type.p2.zoomclick-d3.icon-minus-text-on-apps-with-both.ci90.1\endcsname{$+$17.1}
\expandafter\def\csname val@rev.desc.d7.uiv.tau-T-by-type.p2.zoomclick-d3.icon-minus-text-on-apps-with-both.ci95-excludes-0\endcsname{0}
\expandafter\def\csname cnt@rev.desc.d7.uiv.tau-T-by-type.p2.zoomclick-d3.icon-minus-text-on-apps-with-both.ci95-excludes-0\endcsname{0}
\expandafter\def\csname pct@rev.desc.d7.uiv.tau-T-by-type.p2.zoomclick-d3.icon-minus-text-on-apps-with-both.ci95-excludes-0\endcsname{0.0}
\expandafter\def\csname ppt@rev.desc.d7.uiv.tau-T-by-type.p2.zoomclick-d3.icon-minus-text-on-apps-with-both.ci95-excludes-0\endcsname{0.0}
\expandafter\def\csname val@rev.desc.d7.uiv.tau-T-by-type.p2.zoomclick-d3.icon-minus-text-on-apps-with-both.ci95.0\endcsname{-0.200}
\expandafter\def\csname cnt@rev.desc.d7.uiv.tau-T-by-type.p2.zoomclick-d3.icon-minus-text-on-apps-with-both.ci95.0\endcsname{0}
\expandafter\def\csname pct@rev.desc.d7.uiv.tau-T-by-type.p2.zoomclick-d3.icon-minus-text-on-apps-with-both.ci95.0\endcsname{$-$20.0}
\expandafter\def\csname ppt@rev.desc.d7.uiv.tau-T-by-type.p2.zoomclick-d3.icon-minus-text-on-apps-with-both.ci95.0\endcsname{$-$20.0}
\expandafter\def\csname val@rev.desc.d7.uiv.tau-T-by-type.p2.zoomclick-d3.icon-minus-text-on-apps-with-both.ci95.1\endcsname{0.214}
\expandafter\def\csname cnt@rev.desc.d7.uiv.tau-T-by-type.p2.zoomclick-d3.icon-minus-text-on-apps-with-both.ci95.1\endcsname{0}
\expandafter\def\csname pct@rev.desc.d7.uiv.tau-T-by-type.p2.zoomclick-d3.icon-minus-text-on-apps-with-both.ci95.1\endcsname{21.4}
\expandafter\def\csname ppt@rev.desc.d7.uiv.tau-T-by-type.p2.zoomclick-d3.icon-minus-text-on-apps-with-both.ci95.1\endcsname{$+$21.4}
\expandafter\def\csname val@rev.desc.d7.uiv.tau-T-by-type.p2.zoomclick-d3.icon-minus-text-on-apps-with-both.mean\endcsname{0.014}
\expandafter\def\csname cnt@rev.desc.d7.uiv.tau-T-by-type.p2.zoomclick-d3.icon-minus-text-on-apps-with-both.mean\endcsname{0}
\expandafter\def\csname pct@rev.desc.d7.uiv.tau-T-by-type.p2.zoomclick-d3.icon-minus-text-on-apps-with-both.mean\endcsname{1.4}
\expandafter\def\csname ppt@rev.desc.d7.uiv.tau-T-by-type.p2.zoomclick-d3.icon-minus-text-on-apps-with-both.mean\endcsname{$+$1.4}
\expandafter\def\csname val@rev.desc.d7.uiv.tau-T-by-type.p2.zoomclick-d3.icon-minus-text-on-apps-with-both.n-apps\endcsname{14}
\expandafter\def\csname cnt@rev.desc.d7.uiv.tau-T-by-type.p2.zoomclick-d3.icon-minus-text-on-apps-with-both.n-apps\endcsname{14}
\expandafter\def\csname pct@rev.desc.d7.uiv.tau-T-by-type.p2.zoomclick-d3.icon-minus-text-on-apps-with-both.n-apps\endcsname{1400.0}
\expandafter\def\csname ppt@rev.desc.d7.uiv.tau-T-by-type.p2.zoomclick-d3.icon-minus-text-on-apps-with-both.n-apps\endcsname{$+$1400.0}
\expandafter\def\csname val@rev.desc.d7.uiv.tau-T-by-type.p2.zoomclick-d3.icon-minus-text-on-apps-with-both.tost1.ci90-inside\endcsname{0}
\expandafter\def\csname cnt@rev.desc.d7.uiv.tau-T-by-type.p2.zoomclick-d3.icon-minus-text-on-apps-with-both.tost1.ci90-inside\endcsname{0}
\expandafter\def\csname pct@rev.desc.d7.uiv.tau-T-by-type.p2.zoomclick-d3.icon-minus-text-on-apps-with-both.tost1.ci90-inside\endcsname{0.0}
\expandafter\def\csname ppt@rev.desc.d7.uiv.tau-T-by-type.p2.zoomclick-d3.icon-minus-text-on-apps-with-both.tost1.ci90-inside\endcsname{0.0}
\expandafter\def\csname val@rev.desc.d7.uiv.tau-T-by-type.p2.zoomclick-d3.icon-minus-text-on-apps-with-both.tost1.equivalent\endcsname{0}
\expandafter\def\csname cnt@rev.desc.d7.uiv.tau-T-by-type.p2.zoomclick-d3.icon-minus-text-on-apps-with-both.tost1.equivalent\endcsname{0}
\expandafter\def\csname pct@rev.desc.d7.uiv.tau-T-by-type.p2.zoomclick-d3.icon-minus-text-on-apps-with-both.tost1.equivalent\endcsname{0.0}
\expandafter\def\csname ppt@rev.desc.d7.uiv.tau-T-by-type.p2.zoomclick-d3.icon-minus-text-on-apps-with-both.tost1.equivalent\endcsname{0.0}
\expandafter\def\csname val@rev.desc.d7.uiv.tau-T-by-type.p2.zoomclick-d3.icon-minus-text-on-apps-with-both.tost1.p\endcsname{0.558}
\expandafter\def\csname cnt@rev.desc.d7.uiv.tau-T-by-type.p2.zoomclick-d3.icon-minus-text-on-apps-with-both.tost1.p\endcsname{1}
\expandafter\def\csname pct@rev.desc.d7.uiv.tau-T-by-type.p2.zoomclick-d3.icon-minus-text-on-apps-with-both.tost1.p\endcsname{55.8}
\expandafter\def\csname ppt@rev.desc.d7.uiv.tau-T-by-type.p2.zoomclick-d3.icon-minus-text-on-apps-with-both.tost1.p\endcsname{$+$55.8}
\expandafter\def\csname val@rev.desc.d7.uiv.tau-T-by-type.p2.zoomclick-d3.icon-minus-text-on-apps-with-both.tost5.ci90-inside\endcsname{0}
\expandafter\def\csname cnt@rev.desc.d7.uiv.tau-T-by-type.p2.zoomclick-d3.icon-minus-text-on-apps-with-both.tost5.ci90-inside\endcsname{0}
\expandafter\def\csname pct@rev.desc.d7.uiv.tau-T-by-type.p2.zoomclick-d3.icon-minus-text-on-apps-with-both.tost5.ci90-inside\endcsname{0.0}
\expandafter\def\csname ppt@rev.desc.d7.uiv.tau-T-by-type.p2.zoomclick-d3.icon-minus-text-on-apps-with-both.tost5.ci90-inside\endcsname{0.0}
\expandafter\def\csname val@rev.desc.d7.uiv.tau-T-by-type.p2.zoomclick-d3.icon-minus-text-on-apps-with-both.tost5.equivalent\endcsname{0}
\expandafter\def\csname cnt@rev.desc.d7.uiv.tau-T-by-type.p2.zoomclick-d3.icon-minus-text-on-apps-with-both.tost5.equivalent\endcsname{0}
\expandafter\def\csname pct@rev.desc.d7.uiv.tau-T-by-type.p2.zoomclick-d3.icon-minus-text-on-apps-with-both.tost5.equivalent\endcsname{0.0}
\expandafter\def\csname ppt@rev.desc.d7.uiv.tau-T-by-type.p2.zoomclick-d3.icon-minus-text-on-apps-with-both.tost5.equivalent\endcsname{0.0}
\expandafter\def\csname val@rev.desc.d7.uiv.tau-T-by-type.p2.zoomclick-d3.icon-minus-text-on-apps-with-both.tost5.p\endcsname{0.365}
\expandafter\def\csname cnt@rev.desc.d7.uiv.tau-T-by-type.p2.zoomclick-d3.icon-minus-text-on-apps-with-both.tost5.p\endcsname{0}
\expandafter\def\csname pct@rev.desc.d7.uiv.tau-T-by-type.p2.zoomclick-d3.icon-minus-text-on-apps-with-both.tost5.p\endcsname{36.5}
\expandafter\def\csname ppt@rev.desc.d7.uiv.tau-T-by-type.p2.zoomclick-d3.icon-minus-text-on-apps-with-both.tost5.p\endcsname{$+$36.5}
\expandafter\def\csname val@rev.desc.d7.uiv.tau-T-by-type.p2.zoomclick-d3.icon.acc-F\endcsname{0.842}
\expandafter\def\csname cnt@rev.desc.d7.uiv.tau-T-by-type.p2.zoomclick-d3.icon.acc-F\endcsname{1}
\expandafter\def\csname pct@rev.desc.d7.uiv.tau-T-by-type.p2.zoomclick-d3.icon.acc-F\endcsname{84.2}
\expandafter\def\csname ppt@rev.desc.d7.uiv.tau-T-by-type.p2.zoomclick-d3.icon.acc-F\endcsname{$+$84.2}
\expandafter\def\csname val@rev.desc.d7.uiv.tau-T-by-type.p2.zoomclick-d3.icon.acc-direct\endcsname{0.851}
\expandafter\def\csname cnt@rev.desc.d7.uiv.tau-T-by-type.p2.zoomclick-d3.icon.acc-direct\endcsname{1}
\expandafter\def\csname pct@rev.desc.d7.uiv.tau-T-by-type.p2.zoomclick-d3.icon.acc-direct\endcsname{85.1}
\expandafter\def\csname ppt@rev.desc.d7.uiv.tau-T-by-type.p2.zoomclick-d3.icon.acc-direct\endcsname{$+$85.1}
\expandafter\def\csname val@rev.desc.d7.uiv.tau-T-by-type.p2.zoomclick-d3.icon.n-apps\endcsname{23}
\expandafter\def\csname cnt@rev.desc.d7.uiv.tau-T-by-type.p2.zoomclick-d3.icon.n-apps\endcsname{23}
\expandafter\def\csname pct@rev.desc.d7.uiv.tau-T-by-type.p2.zoomclick-d3.icon.n-apps\endcsname{2300.0}
\expandafter\def\csname ppt@rev.desc.d7.uiv.tau-T-by-type.p2.zoomclick-d3.icon.n-apps\endcsname{$+$2300.0}
\expandafter\def\csname val@rev.desc.d7.uiv.tau-T-by-type.p2.zoomclick-d3.icon.n-tasks\endcsname{114}
\expandafter\def\csname cnt@rev.desc.d7.uiv.tau-T-by-type.p2.zoomclick-d3.icon.n-tasks\endcsname{114}
\expandafter\def\csname pct@rev.desc.d7.uiv.tau-T-by-type.p2.zoomclick-d3.icon.n-tasks\endcsname{11400.0}
\expandafter\def\csname ppt@rev.desc.d7.uiv.tau-T-by-type.p2.zoomclick-d3.icon.n-tasks\endcsname{$+$11400.0}
\expandafter\def\csname val@rev.desc.d7.uiv.tau-T-by-type.p2.zoomclick-d3.icon.tau.ci90.0\endcsname{-0.036}
\expandafter\def\csname cnt@rev.desc.d7.uiv.tau-T-by-type.p2.zoomclick-d3.icon.tau.ci90.0\endcsname{0}
\expandafter\def\csname pct@rev.desc.d7.uiv.tau-T-by-type.p2.zoomclick-d3.icon.tau.ci90.0\endcsname{$-$3.6}
\expandafter\def\csname ppt@rev.desc.d7.uiv.tau-T-by-type.p2.zoomclick-d3.icon.tau.ci90.0\endcsname{$-$3.6}
\expandafter\def\csname val@rev.desc.d7.uiv.tau-T-by-type.p2.zoomclick-d3.icon.tau.ci90.1\endcsname{0.025}
\expandafter\def\csname cnt@rev.desc.d7.uiv.tau-T-by-type.p2.zoomclick-d3.icon.tau.ci90.1\endcsname{0}
\expandafter\def\csname pct@rev.desc.d7.uiv.tau-T-by-type.p2.zoomclick-d3.icon.tau.ci90.1\endcsname{2.5}
\expandafter\def\csname ppt@rev.desc.d7.uiv.tau-T-by-type.p2.zoomclick-d3.icon.tau.ci90.1\endcsname{$+$2.5}
\expandafter\def\csname val@rev.desc.d7.uiv.tau-T-by-type.p2.zoomclick-d3.icon.tau.ci95-excludes-0\endcsname{0}
\expandafter\def\csname cnt@rev.desc.d7.uiv.tau-T-by-type.p2.zoomclick-d3.icon.tau.ci95-excludes-0\endcsname{0}
\expandafter\def\csname pct@rev.desc.d7.uiv.tau-T-by-type.p2.zoomclick-d3.icon.tau.ci95-excludes-0\endcsname{0.0}
\expandafter\def\csname ppt@rev.desc.d7.uiv.tau-T-by-type.p2.zoomclick-d3.icon.tau.ci95-excludes-0\endcsname{0.0}
\expandafter\def\csname val@rev.desc.d7.uiv.tau-T-by-type.p2.zoomclick-d3.icon.tau.ci95.0\endcsname{-0.043}
\expandafter\def\csname cnt@rev.desc.d7.uiv.tau-T-by-type.p2.zoomclick-d3.icon.tau.ci95.0\endcsname{0}
\expandafter\def\csname pct@rev.desc.d7.uiv.tau-T-by-type.p2.zoomclick-d3.icon.tau.ci95.0\endcsname{$-$4.3}
\expandafter\def\csname ppt@rev.desc.d7.uiv.tau-T-by-type.p2.zoomclick-d3.icon.tau.ci95.0\endcsname{$-$4.3}
\expandafter\def\csname val@rev.desc.d7.uiv.tau-T-by-type.p2.zoomclick-d3.icon.tau.ci95.1\endcsname{0.030}
\expandafter\def\csname cnt@rev.desc.d7.uiv.tau-T-by-type.p2.zoomclick-d3.icon.tau.ci95.1\endcsname{0}
\expandafter\def\csname pct@rev.desc.d7.uiv.tau-T-by-type.p2.zoomclick-d3.icon.tau.ci95.1\endcsname{3.0}
\expandafter\def\csname ppt@rev.desc.d7.uiv.tau-T-by-type.p2.zoomclick-d3.icon.tau.ci95.1\endcsname{$+$3.0}
\expandafter\def\csname val@rev.desc.d7.uiv.tau-T-by-type.p2.zoomclick-d3.icon.tau.mean\endcsname{-0.006}
\expandafter\def\csname cnt@rev.desc.d7.uiv.tau-T-by-type.p2.zoomclick-d3.icon.tau.mean\endcsname{0}
\expandafter\def\csname pct@rev.desc.d7.uiv.tau-T-by-type.p2.zoomclick-d3.icon.tau.mean\endcsname{$-$0.6}
\expandafter\def\csname ppt@rev.desc.d7.uiv.tau-T-by-type.p2.zoomclick-d3.icon.tau.mean\endcsname{$-$0.6}
\expandafter\def\csname val@rev.desc.d7.uiv.tau-T-by-type.p2.zoomclick-d3.icon.tau.n-apps\endcsname{23}
\expandafter\def\csname cnt@rev.desc.d7.uiv.tau-T-by-type.p2.zoomclick-d3.icon.tau.n-apps\endcsname{23}
\expandafter\def\csname pct@rev.desc.d7.uiv.tau-T-by-type.p2.zoomclick-d3.icon.tau.n-apps\endcsname{2300.0}
\expandafter\def\csname ppt@rev.desc.d7.uiv.tau-T-by-type.p2.zoomclick-d3.icon.tau.n-apps\endcsname{$+$2300.0}
\expandafter\def\csname val@rev.desc.d7.uiv.tau-T-by-type.p2.zoomclick-d3.icon.tau.tost1.ci90-inside\endcsname{0}
\expandafter\def\csname cnt@rev.desc.d7.uiv.tau-T-by-type.p2.zoomclick-d3.icon.tau.tost1.ci90-inside\endcsname{0}
\expandafter\def\csname pct@rev.desc.d7.uiv.tau-T-by-type.p2.zoomclick-d3.icon.tau.tost1.ci90-inside\endcsname{0.0}
\expandafter\def\csname ppt@rev.desc.d7.uiv.tau-T-by-type.p2.zoomclick-d3.icon.tau.tost1.ci90-inside\endcsname{0.0}
\expandafter\def\csname val@rev.desc.d7.uiv.tau-T-by-type.p2.zoomclick-d3.icon.tau.tost1.equivalent\endcsname{0}
\expandafter\def\csname cnt@rev.desc.d7.uiv.tau-T-by-type.p2.zoomclick-d3.icon.tau.tost1.equivalent\endcsname{0}
\expandafter\def\csname pct@rev.desc.d7.uiv.tau-T-by-type.p2.zoomclick-d3.icon.tau.tost1.equivalent\endcsname{0.0}
\expandafter\def\csname ppt@rev.desc.d7.uiv.tau-T-by-type.p2.zoomclick-d3.icon.tau.tost1.equivalent\endcsname{0.0}
\expandafter\def\csname val@rev.desc.d7.uiv.tau-T-by-type.p2.zoomclick-d3.icon.tau.tost1.p\endcsname{0.391}
\expandafter\def\csname cnt@rev.desc.d7.uiv.tau-T-by-type.p2.zoomclick-d3.icon.tau.tost1.p\endcsname{0}
\expandafter\def\csname pct@rev.desc.d7.uiv.tau-T-by-type.p2.zoomclick-d3.icon.tau.tost1.p\endcsname{39.1}
\expandafter\def\csname ppt@rev.desc.d7.uiv.tau-T-by-type.p2.zoomclick-d3.icon.tau.tost1.p\endcsname{$+$39.1}
\expandafter\def\csname val@rev.desc.d7.uiv.tau-T-by-type.p2.zoomclick-d3.icon.tau.tost5.ci90-inside\endcsname{1}
\expandafter\def\csname cnt@rev.desc.d7.uiv.tau-T-by-type.p2.zoomclick-d3.icon.tau.tost5.ci90-inside\endcsname{1}
\expandafter\def\csname pct@rev.desc.d7.uiv.tau-T-by-type.p2.zoomclick-d3.icon.tau.tost5.ci90-inside\endcsname{100.0}
\expandafter\def\csname ppt@rev.desc.d7.uiv.tau-T-by-type.p2.zoomclick-d3.icon.tau.tost5.ci90-inside\endcsname{$+$100.0}
\expandafter\def\csname val@rev.desc.d7.uiv.tau-T-by-type.p2.zoomclick-d3.icon.tau.tost5.equivalent\endcsname{1}
\expandafter\def\csname cnt@rev.desc.d7.uiv.tau-T-by-type.p2.zoomclick-d3.icon.tau.tost5.equivalent\endcsname{1}
\expandafter\def\csname pct@rev.desc.d7.uiv.tau-T-by-type.p2.zoomclick-d3.icon.tau.tost5.equivalent\endcsname{100.0}
\expandafter\def\csname ppt@rev.desc.d7.uiv.tau-T-by-type.p2.zoomclick-d3.icon.tau.tost5.equivalent\endcsname{$+$100.0}
\expandafter\def\csname val@rev.desc.d7.uiv.tau-T-by-type.p2.zoomclick-d3.icon.tau.tost5.p\endcsname{0.019}
\expandafter\def\csname cnt@rev.desc.d7.uiv.tau-T-by-type.p2.zoomclick-d3.icon.tau.tost5.p\endcsname{0}
\expandafter\def\csname pct@rev.desc.d7.uiv.tau-T-by-type.p2.zoomclick-d3.icon.tau.tost5.p\endcsname{1.9}
\expandafter\def\csname ppt@rev.desc.d7.uiv.tau-T-by-type.p2.zoomclick-d3.icon.tau.tost5.p\endcsname{$+$1.9}
\expandafter\def\csname val@rev.desc.d7.uiv.tau-T-by-type.p2.zoomclick-d3.text.acc-F\endcsname{0.800}
\expandafter\def\csname cnt@rev.desc.d7.uiv.tau-T-by-type.p2.zoomclick-d3.text.acc-F\endcsname{1}
\expandafter\def\csname pct@rev.desc.d7.uiv.tau-T-by-type.p2.zoomclick-d3.text.acc-F\endcsname{80.0}
\expandafter\def\csname ppt@rev.desc.d7.uiv.tau-T-by-type.p2.zoomclick-d3.text.acc-F\endcsname{$+$80.0}
\expandafter\def\csname val@rev.desc.d7.uiv.tau-T-by-type.p2.zoomclick-d3.text.acc-direct\endcsname{0.800}
\expandafter\def\csname cnt@rev.desc.d7.uiv.tau-T-by-type.p2.zoomclick-d3.text.acc-direct\endcsname{1}
\expandafter\def\csname pct@rev.desc.d7.uiv.tau-T-by-type.p2.zoomclick-d3.text.acc-direct\endcsname{80.0}
\expandafter\def\csname ppt@rev.desc.d7.uiv.tau-T-by-type.p2.zoomclick-d3.text.acc-direct\endcsname{$+$80.0}
\expandafter\def\csname val@rev.desc.d7.uiv.tau-T-by-type.p2.zoomclick-d3.text.n-apps\endcsname{15}
\expandafter\def\csname cnt@rev.desc.d7.uiv.tau-T-by-type.p2.zoomclick-d3.text.n-apps\endcsname{15}
\expandafter\def\csname pct@rev.desc.d7.uiv.tau-T-by-type.p2.zoomclick-d3.text.n-apps\endcsname{1500.0}
\expandafter\def\csname ppt@rev.desc.d7.uiv.tau-T-by-type.p2.zoomclick-d3.text.n-apps\endcsname{$+$1500.0}
\expandafter\def\csname val@rev.desc.d7.uiv.tau-T-by-type.p2.zoomclick-d3.text.n-tasks\endcsname{30}
\expandafter\def\csname cnt@rev.desc.d7.uiv.tau-T-by-type.p2.zoomclick-d3.text.n-tasks\endcsname{30}
\expandafter\def\csname pct@rev.desc.d7.uiv.tau-T-by-type.p2.zoomclick-d3.text.n-tasks\endcsname{3000.0}
\expandafter\def\csname ppt@rev.desc.d7.uiv.tau-T-by-type.p2.zoomclick-d3.text.n-tasks\endcsname{$+$3000.0}
\expandafter\def\csname val@rev.desc.d7.uiv.tau-T-by-type.p2.zoomclick-d3.text.tau.ci90.0\endcsname{-0.133}
\expandafter\def\csname cnt@rev.desc.d7.uiv.tau-T-by-type.p2.zoomclick-d3.text.tau.ci90.0\endcsname{0}
\expandafter\def\csname pct@rev.desc.d7.uiv.tau-T-by-type.p2.zoomclick-d3.text.tau.ci90.0\endcsname{$-$13.3}
\expandafter\def\csname ppt@rev.desc.d7.uiv.tau-T-by-type.p2.zoomclick-d3.text.tau.ci90.0\endcsname{$-$13.3}
\expandafter\def\csname val@rev.desc.d7.uiv.tau-T-by-type.p2.zoomclick-d3.text.tau.ci90.1\endcsname{0.133}
\expandafter\def\csname cnt@rev.desc.d7.uiv.tau-T-by-type.p2.zoomclick-d3.text.tau.ci90.1\endcsname{0}
\expandafter\def\csname pct@rev.desc.d7.uiv.tau-T-by-type.p2.zoomclick-d3.text.tau.ci90.1\endcsname{13.3}
\expandafter\def\csname ppt@rev.desc.d7.uiv.tau-T-by-type.p2.zoomclick-d3.text.tau.ci90.1\endcsname{$+$13.3}
\expandafter\def\csname val@rev.desc.d7.uiv.tau-T-by-type.p2.zoomclick-d3.text.tau.ci95-excludes-0\endcsname{0}
\expandafter\def\csname cnt@rev.desc.d7.uiv.tau-T-by-type.p2.zoomclick-d3.text.tau.ci95-excludes-0\endcsname{0}
\expandafter\def\csname pct@rev.desc.d7.uiv.tau-T-by-type.p2.zoomclick-d3.text.tau.ci95-excludes-0\endcsname{0.0}
\expandafter\def\csname ppt@rev.desc.d7.uiv.tau-T-by-type.p2.zoomclick-d3.text.tau.ci95-excludes-0\endcsname{0.0}
\expandafter\def\csname val@rev.desc.d7.uiv.tau-T-by-type.p2.zoomclick-d3.text.tau.ci95.0\endcsname{-0.200}
\expandafter\def\csname cnt@rev.desc.d7.uiv.tau-T-by-type.p2.zoomclick-d3.text.tau.ci95.0\endcsname{0}
\expandafter\def\csname pct@rev.desc.d7.uiv.tau-T-by-type.p2.zoomclick-d3.text.tau.ci95.0\endcsname{$-$20.0}
\expandafter\def\csname ppt@rev.desc.d7.uiv.tau-T-by-type.p2.zoomclick-d3.text.tau.ci95.0\endcsname{$-$20.0}
\expandafter\def\csname val@rev.desc.d7.uiv.tau-T-by-type.p2.zoomclick-d3.text.tau.ci95.1\endcsname{0.200}
\expandafter\def\csname cnt@rev.desc.d7.uiv.tau-T-by-type.p2.zoomclick-d3.text.tau.ci95.1\endcsname{0}
\expandafter\def\csname pct@rev.desc.d7.uiv.tau-T-by-type.p2.zoomclick-d3.text.tau.ci95.1\endcsname{20.0}
\expandafter\def\csname ppt@rev.desc.d7.uiv.tau-T-by-type.p2.zoomclick-d3.text.tau.ci95.1\endcsname{$+$20.0}
\expandafter\def\csname val@rev.desc.d7.uiv.tau-T-by-type.p2.zoomclick-d3.text.tau.mean\endcsname{0.000}
\expandafter\def\csname cnt@rev.desc.d7.uiv.tau-T-by-type.p2.zoomclick-d3.text.tau.mean\endcsname{0}
\expandafter\def\csname pct@rev.desc.d7.uiv.tau-T-by-type.p2.zoomclick-d3.text.tau.mean\endcsname{0.0}
\expandafter\def\csname ppt@rev.desc.d7.uiv.tau-T-by-type.p2.zoomclick-d3.text.tau.mean\endcsname{0.0}
\expandafter\def\csname val@rev.desc.d7.uiv.tau-T-by-type.p2.zoomclick-d3.text.tau.n-apps\endcsname{15}
\expandafter\def\csname cnt@rev.desc.d7.uiv.tau-T-by-type.p2.zoomclick-d3.text.tau.n-apps\endcsname{15}
\expandafter\def\csname pct@rev.desc.d7.uiv.tau-T-by-type.p2.zoomclick-d3.text.tau.n-apps\endcsname{1500.0}
\expandafter\def\csname ppt@rev.desc.d7.uiv.tau-T-by-type.p2.zoomclick-d3.text.tau.n-apps\endcsname{$+$1500.0}
\expandafter\def\csname val@rev.desc.d7.uiv.tau-T-by-type.p2.zoomclick-d3.text.tau.tost1.ci90-inside\endcsname{0}
\expandafter\def\csname cnt@rev.desc.d7.uiv.tau-T-by-type.p2.zoomclick-d3.text.tau.tost1.ci90-inside\endcsname{0}
\expandafter\def\csname pct@rev.desc.d7.uiv.tau-T-by-type.p2.zoomclick-d3.text.tau.tost1.ci90-inside\endcsname{0.0}
\expandafter\def\csname ppt@rev.desc.d7.uiv.tau-T-by-type.p2.zoomclick-d3.text.tau.tost1.ci90-inside\endcsname{0.0}
\expandafter\def\csname val@rev.desc.d7.uiv.tau-T-by-type.p2.zoomclick-d3.text.tau.tost1.equivalent\endcsname{0}
\expandafter\def\csname cnt@rev.desc.d7.uiv.tau-T-by-type.p2.zoomclick-d3.text.tau.tost1.equivalent\endcsname{0}
\expandafter\def\csname pct@rev.desc.d7.uiv.tau-T-by-type.p2.zoomclick-d3.text.tau.tost1.equivalent\endcsname{0.0}
\expandafter\def\csname ppt@rev.desc.d7.uiv.tau-T-by-type.p2.zoomclick-d3.text.tau.tost1.equivalent\endcsname{0.0}
\expandafter\def\csname val@rev.desc.d7.uiv.tau-T-by-type.p2.zoomclick-d3.text.tau.tost1.p\endcsname{0.351}
\expandafter\def\csname cnt@rev.desc.d7.uiv.tau-T-by-type.p2.zoomclick-d3.text.tau.tost1.p\endcsname{0}
\expandafter\def\csname pct@rev.desc.d7.uiv.tau-T-by-type.p2.zoomclick-d3.text.tau.tost1.p\endcsname{35.1}
\expandafter\def\csname ppt@rev.desc.d7.uiv.tau-T-by-type.p2.zoomclick-d3.text.tau.tost1.p\endcsname{$+$35.1}
\expandafter\def\csname val@rev.desc.d7.uiv.tau-T-by-type.p2.zoomclick-d3.text.tau.tost5.ci90-inside\endcsname{0}
\expandafter\def\csname cnt@rev.desc.d7.uiv.tau-T-by-type.p2.zoomclick-d3.text.tau.tost5.ci90-inside\endcsname{0}
\expandafter\def\csname pct@rev.desc.d7.uiv.tau-T-by-type.p2.zoomclick-d3.text.tau.tost5.ci90-inside\endcsname{0.0}
\expandafter\def\csname ppt@rev.desc.d7.uiv.tau-T-by-type.p2.zoomclick-d3.text.tau.tost5.ci90-inside\endcsname{0.0}
\expandafter\def\csname val@rev.desc.d7.uiv.tau-T-by-type.p2.zoomclick-d3.text.tau.tost5.equivalent\endcsname{0}
\expandafter\def\csname cnt@rev.desc.d7.uiv.tau-T-by-type.p2.zoomclick-d3.text.tau.tost5.equivalent\endcsname{0}
\expandafter\def\csname pct@rev.desc.d7.uiv.tau-T-by-type.p2.zoomclick-d3.text.tau.tost5.equivalent\endcsname{0.0}
\expandafter\def\csname ppt@rev.desc.d7.uiv.tau-T-by-type.p2.zoomclick-d3.text.tau.tost5.equivalent\endcsname{0.0}
\expandafter\def\csname val@rev.desc.d7.uiv.tau-T-by-type.p2.zoomclick-d3.text.tau.tost5.p\endcsname{0.351}
\expandafter\def\csname cnt@rev.desc.d7.uiv.tau-T-by-type.p2.zoomclick-d3.text.tau.tost5.p\endcsname{0}
\expandafter\def\csname pct@rev.desc.d7.uiv.tau-T-by-type.p2.zoomclick-d3.text.tau.tost5.p\endcsname{35.1}
\expandafter\def\csname ppt@rev.desc.d7.uiv.tau-T-by-type.p2.zoomclick-d3.text.tau.tost5.p\endcsname{$+$35.1}
\expandafter\def\csname val@rev.desc.d8.n-boundary\endcsname{158}
\expandafter\def\csname cnt@rev.desc.d8.n-boundary\endcsname{158}
\expandafter\def\csname pct@rev.desc.d8.n-boundary\endcsname{15800.0}
\expandafter\def\csname ppt@rev.desc.d8.n-boundary\endcsname{$+$15800.0}
\expandafter\def\csname val@rev.desc.d8.n-display-equals-half-even\endcsname{158}
\expandafter\def\csname cnt@rev.desc.d8.n-display-equals-half-even\endcsname{158}
\expandafter\def\csname pct@rev.desc.d8.n-display-equals-half-even\endcsname{15800.0}
\expandafter\def\csname ppt@rev.desc.d8.n-display-equals-half-even\endcsname{$+$15800.0}
\expandafter\def\csname val@rev.desc.d8.n-display-equals-half-up\endcsname{87}
\expandafter\def\csname cnt@rev.desc.d8.n-display-equals-half-up\endcsname{87}
\expandafter\def\csname pct@rev.desc.d8.n-display-equals-half-up\endcsname{8700.0}
\expandafter\def\csname ppt@rev.desc.d8.n-display-equals-half-up\endcsname{$+$8700.0}
\expandafter\def\csname val@rev.desc.d8.n-exact-binary-ties\endcsname{128}
\expandafter\def\csname cnt@rev.desc.d8.n-exact-binary-ties\endcsname{128}
\expandafter\def\csname pct@rev.desc.d8.n-exact-binary-ties\endcsname{12800.0}
\expandafter\def\csname ppt@rev.desc.d8.n-exact-binary-ties\endcsname{$+$12800.0}
\expandafter\def\csname val@rev.desc.d8.n-ppt-boundary-rounded-down\endcsname{89}
\expandafter\def\csname cnt@rev.desc.d8.n-ppt-boundary-rounded-down\endcsname{89}
\expandafter\def\csname pct@rev.desc.d8.n-ppt-boundary-rounded-down\endcsname{8900.0}
\expandafter\def\csname ppt@rev.desc.d8.n-ppt-boundary-rounded-down\endcsname{$+$8900.0}
\expandafter\def\csname val@rev.desc.d8.n-ppt-boundary-rounded-up\endcsname{69}
\expandafter\def\csname cnt@rev.desc.d8.n-ppt-boundary-rounded-up\endcsname{69}
\expandafter\def\csname pct@rev.desc.d8.n-ppt-boundary-rounded-up\endcsname{6900.0}
\expandafter\def\csname ppt@rev.desc.d8.n-ppt-boundary-rounded-up\endcsname{$+$6900.0}
\expandafter\def\csname val@rev.desc.d9.sspro.direct.mean-mpix\endcsname{0.584}
\expandafter\def\csname cnt@rev.desc.d9.sspro.direct.mean-mpix\endcsname{1}
\expandafter\def\csname pct@rev.desc.d9.sspro.direct.mean-mpix\endcsname{58.4}
\expandafter\def\csname ppt@rev.desc.d9.sspro.direct.mean-mpix\endcsname{$+$58.4}
\expandafter\def\csname val@rev.desc.d9.sspro.direct.mean-pixels\endcsname{584246.044}
\expandafter\def\csname cnt@rev.desc.d9.sspro.direct.mean-pixels\endcsname{584{,}246}
\expandafter\def\csname pct@rev.desc.d9.sspro.direct.mean-pixels\endcsname{58424604.4}
\expandafter\def\csname ppt@rev.desc.d9.sspro.direct.mean-pixels\endcsname{$+$58424604.4}
\expandafter\def\csname val@rev.desc.d9.sspro.direct.mean-reader-mpix\endcsname{0.000}
\expandafter\def\csname cnt@rev.desc.d9.sspro.direct.mean-reader-mpix\endcsname{0}
\expandafter\def\csname pct@rev.desc.d9.sspro.direct.mean-reader-mpix\endcsname{0.0}
\expandafter\def\csname ppt@rev.desc.d9.sspro.direct.mean-reader-mpix\endcsname{0.0}
\expandafter\def\csname val@rev.desc.d9.sspro.direct.mean-requests\endcsname{1.000}
\expandafter\def\csname cnt@rev.desc.d9.sspro.direct.mean-requests\endcsname{1}
\expandafter\def\csname pct@rev.desc.d9.sspro.direct.mean-requests\endcsname{100.0}
\expandafter\def\csname ppt@rev.desc.d9.sspro.direct.mean-requests\endcsname{$+$100.0}
\expandafter\def\csname val@rev.desc.d9.sspro.direct.n-jobs\endcsname{1080}
\expandafter\def\csname cnt@rev.desc.d9.sspro.direct.n-jobs\endcsname{1{,}080}
\expandafter\def\csname pct@rev.desc.d9.sspro.direct.n-jobs\endcsname{108000.0}
\expandafter\def\csname ppt@rev.desc.d9.sspro.direct.n-jobs\endcsname{$+$108000.0}
\expandafter\def\csname val@rev.desc.d9.sspro.direct.n-requests-match-scores\endcsname{1}
\expandafter\def\csname cnt@rev.desc.d9.sspro.direct.n-requests-match-scores\endcsname{1}
\expandafter\def\csname pct@rev.desc.d9.sspro.direct.n-requests-match-scores\endcsname{100.0}
\expandafter\def\csname ppt@rev.desc.d9.sspro.direct.n-requests-match-scores\endcsname{$+$100.0}
\expandafter\def\csname val@rev.desc.d9.sspro.direct.ratio-to-direct\endcsname{1.000}
\expandafter\def\csname cnt@rev.desc.d9.sspro.direct.ratio-to-direct\endcsname{1}
\expandafter\def\csname pct@rev.desc.d9.sspro.direct.ratio-to-direct\endcsname{100.0}
\expandafter\def\csname ppt@rev.desc.d9.sspro.direct.ratio-to-direct\endcsname{$+$100.0}
\expandafter\def\csname val@rev.desc.d9.sspro.natural.center-d1.haiku.mean-mpix\endcsname{1.221}
\expandafter\def\csname cnt@rev.desc.d9.sspro.natural.center-d1.haiku.mean-mpix\endcsname{1}
\expandafter\def\csname pct@rev.desc.d9.sspro.natural.center-d1.haiku.mean-mpix\endcsname{122.1}
\expandafter\def\csname ppt@rev.desc.d9.sspro.natural.center-d1.haiku.mean-mpix\endcsname{$+$122.1}
\expandafter\def\csname val@rev.desc.d9.sspro.natural.center-d1.haiku.mean-pixels\endcsname{1220851.200}
\expandafter\def\csname cnt@rev.desc.d9.sspro.natural.center-d1.haiku.mean-pixels\endcsname{1{,}220{,}851}
\expandafter\def\csname pct@rev.desc.d9.sspro.natural.center-d1.haiku.mean-pixels\endcsname{122085120.0}
\expandafter\def\csname ppt@rev.desc.d9.sspro.natural.center-d1.haiku.mean-pixels\endcsname{$+$122085120.0}
\expandafter\def\csname val@rev.desc.d9.sspro.natural.center-d1.haiku.mean-reader-mpix\endcsname{0.610}
\expandafter\def\csname cnt@rev.desc.d9.sspro.natural.center-d1.haiku.mean-reader-mpix\endcsname{1}
\expandafter\def\csname pct@rev.desc.d9.sspro.natural.center-d1.haiku.mean-reader-mpix\endcsname{61.0}
\expandafter\def\csname ppt@rev.desc.d9.sspro.natural.center-d1.haiku.mean-reader-mpix\endcsname{$+$61.0}
\expandafter\def\csname val@rev.desc.d9.sspro.natural.center-d1.haiku.mean-requests\endcsname{2.000}
\expandafter\def\csname cnt@rev.desc.d9.sspro.natural.center-d1.haiku.mean-requests\endcsname{2}
\expandafter\def\csname pct@rev.desc.d9.sspro.natural.center-d1.haiku.mean-requests\endcsname{200.0}
\expandafter\def\csname ppt@rev.desc.d9.sspro.natural.center-d1.haiku.mean-requests\endcsname{$+$200.0}
\expandafter\def\csname val@rev.desc.d9.sspro.natural.center-d1.haiku.n-jobs\endcsname{80}
\expandafter\def\csname cnt@rev.desc.d9.sspro.natural.center-d1.haiku.n-jobs\endcsname{80}
\expandafter\def\csname pct@rev.desc.d9.sspro.natural.center-d1.haiku.n-jobs\endcsname{8000.0}
\expandafter\def\csname ppt@rev.desc.d9.sspro.natural.center-d1.haiku.n-jobs\endcsname{$+$8000.0}
\expandafter\def\csname val@rev.desc.d9.sspro.natural.center-d1.haiku.n-requests-match-scores\endcsname{1}
\expandafter\def\csname cnt@rev.desc.d9.sspro.natural.center-d1.haiku.n-requests-match-scores\endcsname{1}
\expandafter\def\csname pct@rev.desc.d9.sspro.natural.center-d1.haiku.n-requests-match-scores\endcsname{100.0}
\expandafter\def\csname ppt@rev.desc.d9.sspro.natural.center-d1.haiku.n-requests-match-scores\endcsname{$+$100.0}
\expandafter\def\csname val@rev.desc.d9.sspro.natural.center-d1.haiku.ratio-to-direct\endcsname{2.090}
\expandafter\def\csname cnt@rev.desc.d9.sspro.natural.center-d1.haiku.ratio-to-direct\endcsname{2}
\expandafter\def\csname pct@rev.desc.d9.sspro.natural.center-d1.haiku.ratio-to-direct\endcsname{209.0}
\expandafter\def\csname ppt@rev.desc.d9.sspro.natural.center-d1.haiku.ratio-to-direct\endcsname{$+$209.0}
\expandafter\def\csname val@rev.desc.d9.sspro.natural.center-d1.mean-mpix\endcsname{1.183}
\expandafter\def\csname cnt@rev.desc.d9.sspro.natural.center-d1.mean-mpix\endcsname{1}
\expandafter\def\csname pct@rev.desc.d9.sspro.natural.center-d1.mean-mpix\endcsname{118.3}
\expandafter\def\csname ppt@rev.desc.d9.sspro.natural.center-d1.mean-mpix\endcsname{$+$118.3}
\expandafter\def\csname val@rev.desc.d9.sspro.natural.center-d1.mean-pixels\endcsname{1183153.778}
\expandafter\def\csname cnt@rev.desc.d9.sspro.natural.center-d1.mean-pixels\endcsname{1{,}183{,}154}
\expandafter\def\csname pct@rev.desc.d9.sspro.natural.center-d1.mean-pixels\endcsname{118315377.8}
\expandafter\def\csname ppt@rev.desc.d9.sspro.natural.center-d1.mean-pixels\endcsname{$+$118315377.8}
\expandafter\def\csname val@rev.desc.d9.sspro.natural.center-d1.mean-reader-mpix\endcsname{0.592}
\expandafter\def\csname cnt@rev.desc.d9.sspro.natural.center-d1.mean-reader-mpix\endcsname{1}
\expandafter\def\csname pct@rev.desc.d9.sspro.natural.center-d1.mean-reader-mpix\endcsname{59.2}
\expandafter\def\csname ppt@rev.desc.d9.sspro.natural.center-d1.mean-reader-mpix\endcsname{$+$59.2}
\expandafter\def\csname val@rev.desc.d9.sspro.natural.center-d1.mean-requests\endcsname{2.000}
\expandafter\def\csname cnt@rev.desc.d9.sspro.natural.center-d1.mean-requests\endcsname{2}
\expandafter\def\csname pct@rev.desc.d9.sspro.natural.center-d1.mean-requests\endcsname{200.0}
\expandafter\def\csname ppt@rev.desc.d9.sspro.natural.center-d1.mean-requests\endcsname{$+$200.0}
\expandafter\def\csname val@rev.desc.d9.sspro.natural.center-d1.n-jobs\endcsname{432}
\expandafter\def\csname cnt@rev.desc.d9.sspro.natural.center-d1.n-jobs\endcsname{432}
\expandafter\def\csname pct@rev.desc.d9.sspro.natural.center-d1.n-jobs\endcsname{43200.0}
\expandafter\def\csname ppt@rev.desc.d9.sspro.natural.center-d1.n-jobs\endcsname{$+$43200.0}
\expandafter\def\csname val@rev.desc.d9.sspro.natural.center-d1.n-requests-match-scores\endcsname{1}
\expandafter\def\csname cnt@rev.desc.d9.sspro.natural.center-d1.n-requests-match-scores\endcsname{1}
\expandafter\def\csname pct@rev.desc.d9.sspro.natural.center-d1.n-requests-match-scores\endcsname{100.0}
\expandafter\def\csname ppt@rev.desc.d9.sspro.natural.center-d1.n-requests-match-scores\endcsname{$+$100.0}
\expandafter\def\csname val@rev.desc.d9.sspro.natural.center-d1.opus.mean-mpix\endcsname{1.175}
\expandafter\def\csname cnt@rev.desc.d9.sspro.natural.center-d1.opus.mean-mpix\endcsname{1}
\expandafter\def\csname pct@rev.desc.d9.sspro.natural.center-d1.opus.mean-mpix\endcsname{117.5}
\expandafter\def\csname ppt@rev.desc.d9.sspro.natural.center-d1.opus.mean-mpix\endcsname{$+$117.5}
\expandafter\def\csname val@rev.desc.d9.sspro.natural.center-d1.opus.mean-pixels\endcsname{1174580.364}
\expandafter\def\csname cnt@rev.desc.d9.sspro.natural.center-d1.opus.mean-pixels\endcsname{1{,}174{,}580}
\expandafter\def\csname pct@rev.desc.d9.sspro.natural.center-d1.opus.mean-pixels\endcsname{117458036.4}
\expandafter\def\csname ppt@rev.desc.d9.sspro.natural.center-d1.opus.mean-pixels\endcsname{$+$117458036.4}
\expandafter\def\csname val@rev.desc.d9.sspro.natural.center-d1.opus.mean-reader-mpix\endcsname{0.587}
\expandafter\def\csname cnt@rev.desc.d9.sspro.natural.center-d1.opus.mean-reader-mpix\endcsname{1}
\expandafter\def\csname pct@rev.desc.d9.sspro.natural.center-d1.opus.mean-reader-mpix\endcsname{58.7}
\expandafter\def\csname ppt@rev.desc.d9.sspro.natural.center-d1.opus.mean-reader-mpix\endcsname{$+$58.7}
\expandafter\def\csname val@rev.desc.d9.sspro.natural.center-d1.opus.mean-requests\endcsname{2.000}
\expandafter\def\csname cnt@rev.desc.d9.sspro.natural.center-d1.opus.mean-requests\endcsname{2}
\expandafter\def\csname pct@rev.desc.d9.sspro.natural.center-d1.opus.mean-requests\endcsname{200.0}
\expandafter\def\csname ppt@rev.desc.d9.sspro.natural.center-d1.opus.mean-requests\endcsname{$+$200.0}
\expandafter\def\csname val@rev.desc.d9.sspro.natural.center-d1.opus.n-jobs\endcsname{176}
\expandafter\def\csname cnt@rev.desc.d9.sspro.natural.center-d1.opus.n-jobs\endcsname{176}
\expandafter\def\csname pct@rev.desc.d9.sspro.natural.center-d1.opus.n-jobs\endcsname{17600.0}
\expandafter\def\csname ppt@rev.desc.d9.sspro.natural.center-d1.opus.n-jobs\endcsname{$+$17600.0}
\expandafter\def\csname val@rev.desc.d9.sspro.natural.center-d1.opus.n-requests-match-scores\endcsname{1}
\expandafter\def\csname cnt@rev.desc.d9.sspro.natural.center-d1.opus.n-requests-match-scores\endcsname{1}
\expandafter\def\csname pct@rev.desc.d9.sspro.natural.center-d1.opus.n-requests-match-scores\endcsname{100.0}
\expandafter\def\csname ppt@rev.desc.d9.sspro.natural.center-d1.opus.n-requests-match-scores\endcsname{$+$100.0}
\expandafter\def\csname val@rev.desc.d9.sspro.natural.center-d1.opus.ratio-to-direct\endcsname{2.010}
\expandafter\def\csname cnt@rev.desc.d9.sspro.natural.center-d1.opus.ratio-to-direct\endcsname{2}
\expandafter\def\csname pct@rev.desc.d9.sspro.natural.center-d1.opus.ratio-to-direct\endcsname{201.0}
\expandafter\def\csname ppt@rev.desc.d9.sspro.natural.center-d1.opus.ratio-to-direct\endcsname{$+$201.0}
\expandafter\def\csname val@rev.desc.d9.sspro.natural.center-d1.ratio-to-direct\endcsname{2.025}
\expandafter\def\csname cnt@rev.desc.d9.sspro.natural.center-d1.ratio-to-direct\endcsname{2}
\expandafter\def\csname pct@rev.desc.d9.sspro.natural.center-d1.ratio-to-direct\endcsname{202.5}
\expandafter\def\csname ppt@rev.desc.d9.sspro.natural.center-d1.ratio-to-direct\endcsname{$+$202.5}
\expandafter\def\csname val@rev.desc.d9.sspro.natural.center-d1.sonnet.mean-mpix\endcsname{1.175}
\expandafter\def\csname cnt@rev.desc.d9.sspro.natural.center-d1.sonnet.mean-mpix\endcsname{1}
\expandafter\def\csname pct@rev.desc.d9.sspro.natural.center-d1.sonnet.mean-mpix\endcsname{117.5}
\expandafter\def\csname ppt@rev.desc.d9.sspro.natural.center-d1.sonnet.mean-mpix\endcsname{$+$117.5}
\expandafter\def\csname val@rev.desc.d9.sspro.natural.center-d1.sonnet.mean-pixels\endcsname{1174592.000}
\expandafter\def\csname cnt@rev.desc.d9.sspro.natural.center-d1.sonnet.mean-pixels\endcsname{1{,}174{,}592}
\expandafter\def\csname pct@rev.desc.d9.sspro.natural.center-d1.sonnet.mean-pixels\endcsname{117459200.0}
\expandafter\def\csname ppt@rev.desc.d9.sspro.natural.center-d1.sonnet.mean-pixels\endcsname{$+$117459200.0}
\expandafter\def\csname val@rev.desc.d9.sspro.natural.center-d1.sonnet.mean-reader-mpix\endcsname{0.587}
\expandafter\def\csname cnt@rev.desc.d9.sspro.natural.center-d1.sonnet.mean-reader-mpix\endcsname{1}
\expandafter\def\csname pct@rev.desc.d9.sspro.natural.center-d1.sonnet.mean-reader-mpix\endcsname{58.7}
\expandafter\def\csname ppt@rev.desc.d9.sspro.natural.center-d1.sonnet.mean-reader-mpix\endcsname{$+$58.7}
\expandafter\def\csname val@rev.desc.d9.sspro.natural.center-d1.sonnet.mean-requests\endcsname{2.000}
\expandafter\def\csname cnt@rev.desc.d9.sspro.natural.center-d1.sonnet.mean-requests\endcsname{2}
\expandafter\def\csname pct@rev.desc.d9.sspro.natural.center-d1.sonnet.mean-requests\endcsname{200.0}
\expandafter\def\csname ppt@rev.desc.d9.sspro.natural.center-d1.sonnet.mean-requests\endcsname{$+$200.0}
\expandafter\def\csname val@rev.desc.d9.sspro.natural.center-d1.sonnet.n-jobs\endcsname{176}
\expandafter\def\csname cnt@rev.desc.d9.sspro.natural.center-d1.sonnet.n-jobs\endcsname{176}
\expandafter\def\csname pct@rev.desc.d9.sspro.natural.center-d1.sonnet.n-jobs\endcsname{17600.0}
\expandafter\def\csname ppt@rev.desc.d9.sspro.natural.center-d1.sonnet.n-jobs\endcsname{$+$17600.0}
\expandafter\def\csname val@rev.desc.d9.sspro.natural.center-d1.sonnet.n-requests-match-scores\endcsname{1}
\expandafter\def\csname cnt@rev.desc.d9.sspro.natural.center-d1.sonnet.n-requests-match-scores\endcsname{1}
\expandafter\def\csname pct@rev.desc.d9.sspro.natural.center-d1.sonnet.n-requests-match-scores\endcsname{100.0}
\expandafter\def\csname ppt@rev.desc.d9.sspro.natural.center-d1.sonnet.n-requests-match-scores\endcsname{$+$100.0}
\expandafter\def\csname val@rev.desc.d9.sspro.natural.center-d1.sonnet.ratio-to-direct\endcsname{2.010}
\expandafter\def\csname cnt@rev.desc.d9.sspro.natural.center-d1.sonnet.ratio-to-direct\endcsname{2}
\expandafter\def\csname pct@rev.desc.d9.sspro.natural.center-d1.sonnet.ratio-to-direct\endcsname{201.0}
\expandafter\def\csname ppt@rev.desc.d9.sspro.natural.center-d1.sonnet.ratio-to-direct\endcsname{$+$201.0}
\expandafter\def\csname val@rev.desc.d9.sspro.natural.context-d1.haiku.mean-mpix\endcsname{1.459}
\expandafter\def\csname cnt@rev.desc.d9.sspro.natural.context-d1.haiku.mean-mpix\endcsname{1}
\expandafter\def\csname pct@rev.desc.d9.sspro.natural.context-d1.haiku.mean-mpix\endcsname{145.9}
\expandafter\def\csname ppt@rev.desc.d9.sspro.natural.context-d1.haiku.mean-mpix\endcsname{$+$145.9}
\expandafter\def\csname val@rev.desc.d9.sspro.natural.context-d1.haiku.mean-pixels\endcsname{1459315.200}
\expandafter\def\csname cnt@rev.desc.d9.sspro.natural.context-d1.haiku.mean-pixels\endcsname{1{,}459{,}315}
\expandafter\def\csname pct@rev.desc.d9.sspro.natural.context-d1.haiku.mean-pixels\endcsname{145931520.0}
\expandafter\def\csname ppt@rev.desc.d9.sspro.natural.context-d1.haiku.mean-pixels\endcsname{$+$145931520.0}
\expandafter\def\csname val@rev.desc.d9.sspro.natural.context-d1.haiku.mean-reader-mpix\endcsname{0.849}
\expandafter\def\csname cnt@rev.desc.d9.sspro.natural.context-d1.haiku.mean-reader-mpix\endcsname{1}
\expandafter\def\csname pct@rev.desc.d9.sspro.natural.context-d1.haiku.mean-reader-mpix\endcsname{84.9}
\expandafter\def\csname ppt@rev.desc.d9.sspro.natural.context-d1.haiku.mean-reader-mpix\endcsname{$+$84.9}
\expandafter\def\csname val@rev.desc.d9.sspro.natural.context-d1.haiku.mean-requests\endcsname{2.000}
\expandafter\def\csname cnt@rev.desc.d9.sspro.natural.context-d1.haiku.mean-requests\endcsname{2}
\expandafter\def\csname pct@rev.desc.d9.sspro.natural.context-d1.haiku.mean-requests\endcsname{200.0}
\expandafter\def\csname ppt@rev.desc.d9.sspro.natural.context-d1.haiku.mean-requests\endcsname{$+$200.0}
\expandafter\def\csname val@rev.desc.d9.sspro.natural.context-d1.haiku.n-jobs\endcsname{80}
\expandafter\def\csname cnt@rev.desc.d9.sspro.natural.context-d1.haiku.n-jobs\endcsname{80}
\expandafter\def\csname pct@rev.desc.d9.sspro.natural.context-d1.haiku.n-jobs\endcsname{8000.0}
\expandafter\def\csname ppt@rev.desc.d9.sspro.natural.context-d1.haiku.n-jobs\endcsname{$+$8000.0}
\expandafter\def\csname val@rev.desc.d9.sspro.natural.context-d1.haiku.n-requests-match-scores\endcsname{1}
\expandafter\def\csname cnt@rev.desc.d9.sspro.natural.context-d1.haiku.n-requests-match-scores\endcsname{1}
\expandafter\def\csname pct@rev.desc.d9.sspro.natural.context-d1.haiku.n-requests-match-scores\endcsname{100.0}
\expandafter\def\csname ppt@rev.desc.d9.sspro.natural.context-d1.haiku.n-requests-match-scores\endcsname{$+$100.0}
\expandafter\def\csname val@rev.desc.d9.sspro.natural.context-d1.haiku.ratio-to-direct\endcsname{2.498}
\expandafter\def\csname cnt@rev.desc.d9.sspro.natural.context-d1.haiku.ratio-to-direct\endcsname{2}
\expandafter\def\csname pct@rev.desc.d9.sspro.natural.context-d1.haiku.ratio-to-direct\endcsname{249.8}
\expandafter\def\csname ppt@rev.desc.d9.sspro.natural.context-d1.haiku.ratio-to-direct\endcsname{$+$249.8}
\expandafter\def\csname val@rev.desc.d9.sspro.natural.context-d1.mean-mpix\endcsname{1.414}
\expandafter\def\csname cnt@rev.desc.d9.sspro.natural.context-d1.mean-mpix\endcsname{1}
\expandafter\def\csname pct@rev.desc.d9.sspro.natural.context-d1.mean-mpix\endcsname{141.4}
\expandafter\def\csname ppt@rev.desc.d9.sspro.natural.context-d1.mean-mpix\endcsname{$+$141.4}
\expandafter\def\csname val@rev.desc.d9.sspro.natural.context-d1.mean-pixels\endcsname{1414238.815}
\expandafter\def\csname cnt@rev.desc.d9.sspro.natural.context-d1.mean-pixels\endcsname{1{,}414{,}239}
\expandafter\def\csname pct@rev.desc.d9.sspro.natural.context-d1.mean-pixels\endcsname{141423881.5}
\expandafter\def\csname ppt@rev.desc.d9.sspro.natural.context-d1.mean-pixels\endcsname{$+$141423881.5}
\expandafter\def\csname val@rev.desc.d9.sspro.natural.context-d1.mean-reader-mpix\endcsname{0.823}
\expandafter\def\csname cnt@rev.desc.d9.sspro.natural.context-d1.mean-reader-mpix\endcsname{1}
\expandafter\def\csname pct@rev.desc.d9.sspro.natural.context-d1.mean-reader-mpix\endcsname{82.3}
\expandafter\def\csname ppt@rev.desc.d9.sspro.natural.context-d1.mean-reader-mpix\endcsname{$+$82.3}
\expandafter\def\csname val@rev.desc.d9.sspro.natural.context-d1.mean-requests\endcsname{2.000}
\expandafter\def\csname cnt@rev.desc.d9.sspro.natural.context-d1.mean-requests\endcsname{2}
\expandafter\def\csname pct@rev.desc.d9.sspro.natural.context-d1.mean-requests\endcsname{200.0}
\expandafter\def\csname ppt@rev.desc.d9.sspro.natural.context-d1.mean-requests\endcsname{$+$200.0}
\expandafter\def\csname val@rev.desc.d9.sspro.natural.context-d1.n-jobs\endcsname{432}
\expandafter\def\csname cnt@rev.desc.d9.sspro.natural.context-d1.n-jobs\endcsname{432}
\expandafter\def\csname pct@rev.desc.d9.sspro.natural.context-d1.n-jobs\endcsname{43200.0}
\expandafter\def\csname ppt@rev.desc.d9.sspro.natural.context-d1.n-jobs\endcsname{$+$43200.0}
\expandafter\def\csname val@rev.desc.d9.sspro.natural.context-d1.n-requests-match-scores\endcsname{1}
\expandafter\def\csname cnt@rev.desc.d9.sspro.natural.context-d1.n-requests-match-scores\endcsname{1}
\expandafter\def\csname pct@rev.desc.d9.sspro.natural.context-d1.n-requests-match-scores\endcsname{100.0}
\expandafter\def\csname ppt@rev.desc.d9.sspro.natural.context-d1.n-requests-match-scores\endcsname{$+$100.0}
\expandafter\def\csname val@rev.desc.d9.sspro.natural.context-d1.opus.mean-mpix\endcsname{1.404}
\expandafter\def\csname cnt@rev.desc.d9.sspro.natural.context-d1.opus.mean-mpix\endcsname{1}
\expandafter\def\csname pct@rev.desc.d9.sspro.natural.context-d1.opus.mean-mpix\endcsname{140.4}
\expandafter\def\csname ppt@rev.desc.d9.sspro.natural.context-d1.opus.mean-mpix\endcsname{$+$140.4}
\expandafter\def\csname val@rev.desc.d9.sspro.natural.context-d1.opus.mean-pixels\endcsname{1403991.273}
\expandafter\def\csname cnt@rev.desc.d9.sspro.natural.context-d1.opus.mean-pixels\endcsname{1{,}403{,}991}
\expandafter\def\csname pct@rev.desc.d9.sspro.natural.context-d1.opus.mean-pixels\endcsname{140399127.3}
\expandafter\def\csname ppt@rev.desc.d9.sspro.natural.context-d1.opus.mean-pixels\endcsname{$+$140399127.3}
\expandafter\def\csname val@rev.desc.d9.sspro.natural.context-d1.opus.mean-reader-mpix\endcsname{0.817}
\expandafter\def\csname cnt@rev.desc.d9.sspro.natural.context-d1.opus.mean-reader-mpix\endcsname{1}
\expandafter\def\csname pct@rev.desc.d9.sspro.natural.context-d1.opus.mean-reader-mpix\endcsname{81.7}
\expandafter\def\csname ppt@rev.desc.d9.sspro.natural.context-d1.opus.mean-reader-mpix\endcsname{$+$81.7}
\expandafter\def\csname val@rev.desc.d9.sspro.natural.context-d1.opus.mean-requests\endcsname{2.000}
\expandafter\def\csname cnt@rev.desc.d9.sspro.natural.context-d1.opus.mean-requests\endcsname{2}
\expandafter\def\csname pct@rev.desc.d9.sspro.natural.context-d1.opus.mean-requests\endcsname{200.0}
\expandafter\def\csname ppt@rev.desc.d9.sspro.natural.context-d1.opus.mean-requests\endcsname{$+$200.0}
\expandafter\def\csname val@rev.desc.d9.sspro.natural.context-d1.opus.n-jobs\endcsname{176}
\expandafter\def\csname cnt@rev.desc.d9.sspro.natural.context-d1.opus.n-jobs\endcsname{176}
\expandafter\def\csname pct@rev.desc.d9.sspro.natural.context-d1.opus.n-jobs\endcsname{17600.0}
\expandafter\def\csname ppt@rev.desc.d9.sspro.natural.context-d1.opus.n-jobs\endcsname{$+$17600.0}
\expandafter\def\csname val@rev.desc.d9.sspro.natural.context-d1.opus.n-requests-match-scores\endcsname{1}
\expandafter\def\csname cnt@rev.desc.d9.sspro.natural.context-d1.opus.n-requests-match-scores\endcsname{1}
\expandafter\def\csname pct@rev.desc.d9.sspro.natural.context-d1.opus.n-requests-match-scores\endcsname{100.0}
\expandafter\def\csname ppt@rev.desc.d9.sspro.natural.context-d1.opus.n-requests-match-scores\endcsname{$+$100.0}
\expandafter\def\csname val@rev.desc.d9.sspro.natural.context-d1.opus.ratio-to-direct\endcsname{2.403}
\expandafter\def\csname cnt@rev.desc.d9.sspro.natural.context-d1.opus.ratio-to-direct\endcsname{2}
\expandafter\def\csname pct@rev.desc.d9.sspro.natural.context-d1.opus.ratio-to-direct\endcsname{240.3}
\expandafter\def\csname ppt@rev.desc.d9.sspro.natural.context-d1.opus.ratio-to-direct\endcsname{$+$240.3}
\expandafter\def\csname val@rev.desc.d9.sspro.natural.context-d1.ratio-to-direct\endcsname{2.421}
\expandafter\def\csname cnt@rev.desc.d9.sspro.natural.context-d1.ratio-to-direct\endcsname{2}
\expandafter\def\csname pct@rev.desc.d9.sspro.natural.context-d1.ratio-to-direct\endcsname{242.1}
\expandafter\def\csname ppt@rev.desc.d9.sspro.natural.context-d1.ratio-to-direct\endcsname{$+$242.1}
\expandafter\def\csname val@rev.desc.d9.sspro.natural.context-d1.sonnet.mean-mpix\endcsname{1.404}
\expandafter\def\csname cnt@rev.desc.d9.sspro.natural.context-d1.sonnet.mean-mpix\endcsname{1}
\expandafter\def\csname pct@rev.desc.d9.sspro.natural.context-d1.sonnet.mean-mpix\endcsname{140.4}
\expandafter\def\csname ppt@rev.desc.d9.sspro.natural.context-d1.sonnet.mean-mpix\endcsname{$+$140.4}
\expandafter\def\csname val@rev.desc.d9.sspro.natural.context-d1.sonnet.mean-pixels\endcsname{1403997.091}
\expandafter\def\csname cnt@rev.desc.d9.sspro.natural.context-d1.sonnet.mean-pixels\endcsname{1{,}403{,}997}
\expandafter\def\csname pct@rev.desc.d9.sspro.natural.context-d1.sonnet.mean-pixels\endcsname{140399709.1}
\expandafter\def\csname ppt@rev.desc.d9.sspro.natural.context-d1.sonnet.mean-pixels\endcsname{$+$140399709.1}
\expandafter\def\csname val@rev.desc.d9.sspro.natural.context-d1.sonnet.mean-reader-mpix\endcsname{0.817}
\expandafter\def\csname cnt@rev.desc.d9.sspro.natural.context-d1.sonnet.mean-reader-mpix\endcsname{1}
\expandafter\def\csname pct@rev.desc.d9.sspro.natural.context-d1.sonnet.mean-reader-mpix\endcsname{81.7}
\expandafter\def\csname ppt@rev.desc.d9.sspro.natural.context-d1.sonnet.mean-reader-mpix\endcsname{$+$81.7}
\expandafter\def\csname val@rev.desc.d9.sspro.natural.context-d1.sonnet.mean-requests\endcsname{2.000}
\expandafter\def\csname cnt@rev.desc.d9.sspro.natural.context-d1.sonnet.mean-requests\endcsname{2}
\expandafter\def\csname pct@rev.desc.d9.sspro.natural.context-d1.sonnet.mean-requests\endcsname{200.0}
\expandafter\def\csname ppt@rev.desc.d9.sspro.natural.context-d1.sonnet.mean-requests\endcsname{$+$200.0}
\expandafter\def\csname val@rev.desc.d9.sspro.natural.context-d1.sonnet.n-jobs\endcsname{176}
\expandafter\def\csname cnt@rev.desc.d9.sspro.natural.context-d1.sonnet.n-jobs\endcsname{176}
\expandafter\def\csname pct@rev.desc.d9.sspro.natural.context-d1.sonnet.n-jobs\endcsname{17600.0}
\expandafter\def\csname ppt@rev.desc.d9.sspro.natural.context-d1.sonnet.n-jobs\endcsname{$+$17600.0}
\expandafter\def\csname val@rev.desc.d9.sspro.natural.context-d1.sonnet.n-requests-match-scores\endcsname{1}
\expandafter\def\csname cnt@rev.desc.d9.sspro.natural.context-d1.sonnet.n-requests-match-scores\endcsname{1}
\expandafter\def\csname pct@rev.desc.d9.sspro.natural.context-d1.sonnet.n-requests-match-scores\endcsname{100.0}
\expandafter\def\csname ppt@rev.desc.d9.sspro.natural.context-d1.sonnet.n-requests-match-scores\endcsname{$+$100.0}
\expandafter\def\csname val@rev.desc.d9.sspro.natural.context-d1.sonnet.ratio-to-direct\endcsname{2.403}
\expandafter\def\csname cnt@rev.desc.d9.sspro.natural.context-d1.sonnet.ratio-to-direct\endcsname{2}
\expandafter\def\csname pct@rev.desc.d9.sspro.natural.context-d1.sonnet.ratio-to-direct\endcsname{240.3}
\expandafter\def\csname ppt@rev.desc.d9.sspro.natural.context-d1.sonnet.ratio-to-direct\endcsname{$+$240.3}
\expandafter\def\csname val@rev.desc.d9.sspro.natural.dual-d2.haiku.mean-mpix\endcsname{1.126}
\expandafter\def\csname cnt@rev.desc.d9.sspro.natural.dual-d2.haiku.mean-mpix\endcsname{1}
\expandafter\def\csname pct@rev.desc.d9.sspro.natural.dual-d2.haiku.mean-mpix\endcsname{112.6}
\expandafter\def\csname ppt@rev.desc.d9.sspro.natural.dual-d2.haiku.mean-mpix\endcsname{$+$112.6}
\expandafter\def\csname val@rev.desc.d9.sspro.natural.dual-d2.haiku.mean-pixels\endcsname{1126039.413}
\expandafter\def\csname cnt@rev.desc.d9.sspro.natural.dual-d2.haiku.mean-pixels\endcsname{1{,}126{,}039}
\expandafter\def\csname pct@rev.desc.d9.sspro.natural.dual-d2.haiku.mean-pixels\endcsname{112603941.3}
\expandafter\def\csname ppt@rev.desc.d9.sspro.natural.dual-d2.haiku.mean-pixels\endcsname{$+$112603941.3}
\expandafter\def\csname val@rev.desc.d9.sspro.natural.dual-d2.haiku.mean-reader-mpix\endcsname{0.333}
\expandafter\def\csname cnt@rev.desc.d9.sspro.natural.dual-d2.haiku.mean-reader-mpix\endcsname{0}
\expandafter\def\csname pct@rev.desc.d9.sspro.natural.dual-d2.haiku.mean-reader-mpix\endcsname{33.3}
\expandafter\def\csname ppt@rev.desc.d9.sspro.natural.dual-d2.haiku.mean-reader-mpix\endcsname{$+$33.3}
\expandafter\def\csname val@rev.desc.d9.sspro.natural.dual-d2.haiku.mean-requests\endcsname{3.013}
\expandafter\def\csname cnt@rev.desc.d9.sspro.natural.dual-d2.haiku.mean-requests\endcsname{3}
\expandafter\def\csname pct@rev.desc.d9.sspro.natural.dual-d2.haiku.mean-requests\endcsname{301.2}
\expandafter\def\csname ppt@rev.desc.d9.sspro.natural.dual-d2.haiku.mean-requests\endcsname{$+$301.2}
\expandafter\def\csname val@rev.desc.d9.sspro.natural.dual-d2.haiku.n-jobs\endcsname{80}
\expandafter\def\csname cnt@rev.desc.d9.sspro.natural.dual-d2.haiku.n-jobs\endcsname{80}
\expandafter\def\csname pct@rev.desc.d9.sspro.natural.dual-d2.haiku.n-jobs\endcsname{8000.0}
\expandafter\def\csname ppt@rev.desc.d9.sspro.natural.dual-d2.haiku.n-jobs\endcsname{$+$8000.0}
\expandafter\def\csname val@rev.desc.d9.sspro.natural.dual-d2.haiku.n-requests-match-scores\endcsname{1}
\expandafter\def\csname cnt@rev.desc.d9.sspro.natural.dual-d2.haiku.n-requests-match-scores\endcsname{1}
\expandafter\def\csname pct@rev.desc.d9.sspro.natural.dual-d2.haiku.n-requests-match-scores\endcsname{100.0}
\expandafter\def\csname ppt@rev.desc.d9.sspro.natural.dual-d2.haiku.n-requests-match-scores\endcsname{$+$100.0}
\expandafter\def\csname val@rev.desc.d9.sspro.natural.dual-d2.haiku.ratio-to-direct\endcsname{1.927}
\expandafter\def\csname cnt@rev.desc.d9.sspro.natural.dual-d2.haiku.ratio-to-direct\endcsname{2}
\expandafter\def\csname pct@rev.desc.d9.sspro.natural.dual-d2.haiku.ratio-to-direct\endcsname{192.7}
\expandafter\def\csname ppt@rev.desc.d9.sspro.natural.dual-d2.haiku.ratio-to-direct\endcsname{$+$192.7}
\expandafter\def\csname val@rev.desc.d9.sspro.natural.dual-d2.mean-mpix\endcsname{1.040}
\expandafter\def\csname cnt@rev.desc.d9.sspro.natural.dual-d2.mean-mpix\endcsname{1}
\expandafter\def\csname pct@rev.desc.d9.sspro.natural.dual-d2.mean-mpix\endcsname{104.0}
\expandafter\def\csname ppt@rev.desc.d9.sspro.natural.dual-d2.mean-mpix\endcsname{$+$104.0}
\expandafter\def\csname val@rev.desc.d9.sspro.natural.dual-d2.mean-pixels\endcsname{1040113.051}
\expandafter\def\csname cnt@rev.desc.d9.sspro.natural.dual-d2.mean-pixels\endcsname{1{,}040{,}113}
\expandafter\def\csname pct@rev.desc.d9.sspro.natural.dual-d2.mean-pixels\endcsname{104011305.1}
\expandafter\def\csname ppt@rev.desc.d9.sspro.natural.dual-d2.mean-pixels\endcsname{$+$104011305.1}
\expandafter\def\csname val@rev.desc.d9.sspro.natural.dual-d2.mean-reader-mpix\endcsname{0.301}
\expandafter\def\csname cnt@rev.desc.d9.sspro.natural.dual-d2.mean-reader-mpix\endcsname{0}
\expandafter\def\csname pct@rev.desc.d9.sspro.natural.dual-d2.mean-reader-mpix\endcsname{30.1}
\expandafter\def\csname ppt@rev.desc.d9.sspro.natural.dual-d2.mean-reader-mpix\endcsname{$+$30.1}
\expandafter\def\csname val@rev.desc.d9.sspro.natural.dual-d2.mean-requests\endcsname{3.009}
\expandafter\def\csname cnt@rev.desc.d9.sspro.natural.dual-d2.mean-requests\endcsname{3}
\expandafter\def\csname pct@rev.desc.d9.sspro.natural.dual-d2.mean-requests\endcsname{300.9}
\expandafter\def\csname ppt@rev.desc.d9.sspro.natural.dual-d2.mean-requests\endcsname{$+$300.9}
\expandafter\def\csname val@rev.desc.d9.sspro.natural.dual-d2.n-jobs\endcsname{432}
\expandafter\def\csname cnt@rev.desc.d9.sspro.natural.dual-d2.n-jobs\endcsname{432}
\expandafter\def\csname pct@rev.desc.d9.sspro.natural.dual-d2.n-jobs\endcsname{43200.0}
\expandafter\def\csname ppt@rev.desc.d9.sspro.natural.dual-d2.n-jobs\endcsname{$+$43200.0}
\expandafter\def\csname val@rev.desc.d9.sspro.natural.dual-d2.n-requests-match-scores\endcsname{1}
\expandafter\def\csname cnt@rev.desc.d9.sspro.natural.dual-d2.n-requests-match-scores\endcsname{1}
\expandafter\def\csname pct@rev.desc.d9.sspro.natural.dual-d2.n-requests-match-scores\endcsname{100.0}
\expandafter\def\csname ppt@rev.desc.d9.sspro.natural.dual-d2.n-requests-match-scores\endcsname{$+$100.0}
\expandafter\def\csname val@rev.desc.d9.sspro.natural.dual-d2.opus.mean-mpix\endcsname{0.970}
\expandafter\def\csname cnt@rev.desc.d9.sspro.natural.dual-d2.opus.mean-mpix\endcsname{1}
\expandafter\def\csname pct@rev.desc.d9.sspro.natural.dual-d2.opus.mean-mpix\endcsname{97.0}
\expandafter\def\csname ppt@rev.desc.d9.sspro.natural.dual-d2.opus.mean-mpix\endcsname{$+$97.0}
\expandafter\def\csname val@rev.desc.d9.sspro.natural.dual-d2.opus.mean-pixels\endcsname{970404.364}
\expandafter\def\csname cnt@rev.desc.d9.sspro.natural.dual-d2.opus.mean-pixels\endcsname{970{,}404}
\expandafter\def\csname pct@rev.desc.d9.sspro.natural.dual-d2.opus.mean-pixels\endcsname{97040436.4}
\expandafter\def\csname ppt@rev.desc.d9.sspro.natural.dual-d2.opus.mean-pixels\endcsname{$+$97040436.4}
\expandafter\def\csname val@rev.desc.d9.sspro.natural.dual-d2.opus.mean-reader-mpix\endcsname{0.284}
\expandafter\def\csname cnt@rev.desc.d9.sspro.natural.dual-d2.opus.mean-reader-mpix\endcsname{0}
\expandafter\def\csname pct@rev.desc.d9.sspro.natural.dual-d2.opus.mean-reader-mpix\endcsname{28.4}
\expandafter\def\csname ppt@rev.desc.d9.sspro.natural.dual-d2.opus.mean-reader-mpix\endcsname{$+$28.4}
\expandafter\def\csname val@rev.desc.d9.sspro.natural.dual-d2.opus.mean-requests\endcsname{3.000}
\expandafter\def\csname cnt@rev.desc.d9.sspro.natural.dual-d2.opus.mean-requests\endcsname{3}
\expandafter\def\csname pct@rev.desc.d9.sspro.natural.dual-d2.opus.mean-requests\endcsname{300.0}
\expandafter\def\csname ppt@rev.desc.d9.sspro.natural.dual-d2.opus.mean-requests\endcsname{$+$300.0}
\expandafter\def\csname val@rev.desc.d9.sspro.natural.dual-d2.opus.n-jobs\endcsname{176}
\expandafter\def\csname cnt@rev.desc.d9.sspro.natural.dual-d2.opus.n-jobs\endcsname{176}
\expandafter\def\csname pct@rev.desc.d9.sspro.natural.dual-d2.opus.n-jobs\endcsname{17600.0}
\expandafter\def\csname ppt@rev.desc.d9.sspro.natural.dual-d2.opus.n-jobs\endcsname{$+$17600.0}
\expandafter\def\csname val@rev.desc.d9.sspro.natural.dual-d2.opus.n-requests-match-scores\endcsname{1}
\expandafter\def\csname cnt@rev.desc.d9.sspro.natural.dual-d2.opus.n-requests-match-scores\endcsname{1}
\expandafter\def\csname pct@rev.desc.d9.sspro.natural.dual-d2.opus.n-requests-match-scores\endcsname{100.0}
\expandafter\def\csname ppt@rev.desc.d9.sspro.natural.dual-d2.opus.n-requests-match-scores\endcsname{$+$100.0}
\expandafter\def\csname val@rev.desc.d9.sspro.natural.dual-d2.opus.ratio-to-direct\endcsname{1.661}
\expandafter\def\csname cnt@rev.desc.d9.sspro.natural.dual-d2.opus.ratio-to-direct\endcsname{2}
\expandafter\def\csname pct@rev.desc.d9.sspro.natural.dual-d2.opus.ratio-to-direct\endcsname{166.1}
\expandafter\def\csname ppt@rev.desc.d9.sspro.natural.dual-d2.opus.ratio-to-direct\endcsname{$+$166.1}
\expandafter\def\csname val@rev.desc.d9.sspro.natural.dual-d2.ratio-to-direct\endcsname{1.780}
\expandafter\def\csname cnt@rev.desc.d9.sspro.natural.dual-d2.ratio-to-direct\endcsname{2}
\expandafter\def\csname pct@rev.desc.d9.sspro.natural.dual-d2.ratio-to-direct\endcsname{178.0}
\expandafter\def\csname ppt@rev.desc.d9.sspro.natural.dual-d2.ratio-to-direct\endcsname{$+$178.0}
\expandafter\def\csname val@rev.desc.d9.sspro.natural.dual-d2.sonnet.mean-mpix\endcsname{1.071}
\expandafter\def\csname cnt@rev.desc.d9.sspro.natural.dual-d2.sonnet.mean-mpix\endcsname{1}
\expandafter\def\csname pct@rev.desc.d9.sspro.natural.dual-d2.sonnet.mean-mpix\endcsname{107.1}
\expandafter\def\csname ppt@rev.desc.d9.sspro.natural.dual-d2.sonnet.mean-mpix\endcsname{$+$107.1}
\expandafter\def\csname val@rev.desc.d9.sspro.natural.dual-d2.sonnet.mean-pixels\endcsname{1070764.301}
\expandafter\def\csname cnt@rev.desc.d9.sspro.natural.dual-d2.sonnet.mean-pixels\endcsname{1{,}070{,}764}
\expandafter\def\csname pct@rev.desc.d9.sspro.natural.dual-d2.sonnet.mean-pixels\endcsname{107076430.1}
\expandafter\def\csname ppt@rev.desc.d9.sspro.natural.dual-d2.sonnet.mean-pixels\endcsname{$+$107076430.1}
\expandafter\def\csname val@rev.desc.d9.sspro.natural.dual-d2.sonnet.mean-reader-mpix\endcsname{0.303}
\expandafter\def\csname cnt@rev.desc.d9.sspro.natural.dual-d2.sonnet.mean-reader-mpix\endcsname{0}
\expandafter\def\csname pct@rev.desc.d9.sspro.natural.dual-d2.sonnet.mean-reader-mpix\endcsname{30.3}
\expandafter\def\csname ppt@rev.desc.d9.sspro.natural.dual-d2.sonnet.mean-reader-mpix\endcsname{$+$30.3}
\expandafter\def\csname val@rev.desc.d9.sspro.natural.dual-d2.sonnet.mean-requests\endcsname{3.017}
\expandafter\def\csname cnt@rev.desc.d9.sspro.natural.dual-d2.sonnet.mean-requests\endcsname{3}
\expandafter\def\csname pct@rev.desc.d9.sspro.natural.dual-d2.sonnet.mean-requests\endcsname{301.7}
\expandafter\def\csname ppt@rev.desc.d9.sspro.natural.dual-d2.sonnet.mean-requests\endcsname{$+$301.7}
\expandafter\def\csname val@rev.desc.d9.sspro.natural.dual-d2.sonnet.n-jobs\endcsname{176}
\expandafter\def\csname cnt@rev.desc.d9.sspro.natural.dual-d2.sonnet.n-jobs\endcsname{176}
\expandafter\def\csname pct@rev.desc.d9.sspro.natural.dual-d2.sonnet.n-jobs\endcsname{17600.0}
\expandafter\def\csname ppt@rev.desc.d9.sspro.natural.dual-d2.sonnet.n-jobs\endcsname{$+$17600.0}
\expandafter\def\csname val@rev.desc.d9.sspro.natural.dual-d2.sonnet.n-requests-match-scores\endcsname{1}
\expandafter\def\csname cnt@rev.desc.d9.sspro.natural.dual-d2.sonnet.n-requests-match-scores\endcsname{1}
\expandafter\def\csname pct@rev.desc.d9.sspro.natural.dual-d2.sonnet.n-requests-match-scores\endcsname{100.0}
\expandafter\def\csname ppt@rev.desc.d9.sspro.natural.dual-d2.sonnet.n-requests-match-scores\endcsname{$+$100.0}
\expandafter\def\csname val@rev.desc.d9.sspro.natural.dual-d2.sonnet.ratio-to-direct\endcsname{1.833}
\expandafter\def\csname cnt@rev.desc.d9.sspro.natural.dual-d2.sonnet.ratio-to-direct\endcsname{2}
\expandafter\def\csname pct@rev.desc.d9.sspro.natural.dual-d2.sonnet.ratio-to-direct\endcsname{183.3}
\expandafter\def\csname ppt@rev.desc.d9.sspro.natural.dual-d2.sonnet.ratio-to-direct\endcsname{$+$183.3}
\expandafter\def\csname val@rev.desc.d9.sspro.natural.grid2-d3.haiku.mean-mpix\endcsname{4.094}
\expandafter\def\csname cnt@rev.desc.d9.sspro.natural.grid2-d3.haiku.mean-mpix\endcsname{4}
\expandafter\def\csname pct@rev.desc.d9.sspro.natural.grid2-d3.haiku.mean-mpix\endcsname{409.4}
\expandafter\def\csname ppt@rev.desc.d9.sspro.natural.grid2-d3.haiku.mean-mpix\endcsname{$+$409.4}
\expandafter\def\csname val@rev.desc.d9.sspro.natural.grid2-d3.haiku.mean-pixels\endcsname{4094071.462}
\expandafter\def\csname cnt@rev.desc.d9.sspro.natural.grid2-d3.haiku.mean-pixels\endcsname{4{,}094{,}071}
\expandafter\def\csname pct@rev.desc.d9.sspro.natural.grid2-d3.haiku.mean-pixels\endcsname{409407146.2}
\expandafter\def\csname ppt@rev.desc.d9.sspro.natural.grid2-d3.haiku.mean-pixels\endcsname{$+$409407146.2}
\expandafter\def\csname val@rev.desc.d9.sspro.natural.grid2-d3.haiku.mean-reader-mpix\endcsname{0.476}
\expandafter\def\csname cnt@rev.desc.d9.sspro.natural.grid2-d3.haiku.mean-reader-mpix\endcsname{0}
\expandafter\def\csname pct@rev.desc.d9.sspro.natural.grid2-d3.haiku.mean-reader-mpix\endcsname{47.6}
\expandafter\def\csname ppt@rev.desc.d9.sspro.natural.grid2-d3.haiku.mean-reader-mpix\endcsname{$+$47.6}
\expandafter\def\csname val@rev.desc.d9.sspro.natural.grid2-d3.haiku.mean-requests\endcsname{3.987}
\expandafter\def\csname cnt@rev.desc.d9.sspro.natural.grid2-d3.haiku.mean-requests\endcsname{4}
\expandafter\def\csname pct@rev.desc.d9.sspro.natural.grid2-d3.haiku.mean-requests\endcsname{398.8}
\expandafter\def\csname ppt@rev.desc.d9.sspro.natural.grid2-d3.haiku.mean-requests\endcsname{$+$398.8}
\expandafter\def\csname val@rev.desc.d9.sspro.natural.grid2-d3.haiku.n-jobs\endcsname{80}
\expandafter\def\csname cnt@rev.desc.d9.sspro.natural.grid2-d3.haiku.n-jobs\endcsname{80}
\expandafter\def\csname pct@rev.desc.d9.sspro.natural.grid2-d3.haiku.n-jobs\endcsname{8000.0}
\expandafter\def\csname ppt@rev.desc.d9.sspro.natural.grid2-d3.haiku.n-jobs\endcsname{$+$8000.0}
\expandafter\def\csname val@rev.desc.d9.sspro.natural.grid2-d3.haiku.n-requests-match-scores\endcsname{1}
\expandafter\def\csname cnt@rev.desc.d9.sspro.natural.grid2-d3.haiku.n-requests-match-scores\endcsname{1}
\expandafter\def\csname pct@rev.desc.d9.sspro.natural.grid2-d3.haiku.n-requests-match-scores\endcsname{100.0}
\expandafter\def\csname ppt@rev.desc.d9.sspro.natural.grid2-d3.haiku.n-requests-match-scores\endcsname{$+$100.0}
\expandafter\def\csname val@rev.desc.d9.sspro.natural.grid2-d3.haiku.ratio-to-direct\endcsname{7.007}
\expandafter\def\csname cnt@rev.desc.d9.sspro.natural.grid2-d3.haiku.ratio-to-direct\endcsname{7}
\expandafter\def\csname pct@rev.desc.d9.sspro.natural.grid2-d3.haiku.ratio-to-direct\endcsname{700.7}
\expandafter\def\csname ppt@rev.desc.d9.sspro.natural.grid2-d3.haiku.ratio-to-direct\endcsname{$+$700.7}
\expandafter\def\csname val@rev.desc.d9.sspro.natural.grid2-d3.mean-mpix\endcsname{3.726}
\expandafter\def\csname cnt@rev.desc.d9.sspro.natural.grid2-d3.mean-mpix\endcsname{4}
\expandafter\def\csname pct@rev.desc.d9.sspro.natural.grid2-d3.mean-mpix\endcsname{372.6}
\expandafter\def\csname ppt@rev.desc.d9.sspro.natural.grid2-d3.mean-mpix\endcsname{$+$372.6}
\expandafter\def\csname val@rev.desc.d9.sspro.natural.grid2-d3.mean-pixels\endcsname{3726191.569}
\expandafter\def\csname cnt@rev.desc.d9.sspro.natural.grid2-d3.mean-pixels\endcsname{3{,}726{,}192}
\expandafter\def\csname pct@rev.desc.d9.sspro.natural.grid2-d3.mean-pixels\endcsname{372619156.9}
\expandafter\def\csname ppt@rev.desc.d9.sspro.natural.grid2-d3.mean-pixels\endcsname{$+$372619156.9}
\expandafter\def\csname val@rev.desc.d9.sspro.natural.grid2-d3.mean-reader-mpix\endcsname{0.461}
\expandafter\def\csname cnt@rev.desc.d9.sspro.natural.grid2-d3.mean-reader-mpix\endcsname{0}
\expandafter\def\csname pct@rev.desc.d9.sspro.natural.grid2-d3.mean-reader-mpix\endcsname{46.1}
\expandafter\def\csname ppt@rev.desc.d9.sspro.natural.grid2-d3.mean-reader-mpix\endcsname{$+$46.1}
\expandafter\def\csname val@rev.desc.d9.sspro.natural.grid2-d3.mean-requests\endcsname{3.852}
\expandafter\def\csname cnt@rev.desc.d9.sspro.natural.grid2-d3.mean-requests\endcsname{4}
\expandafter\def\csname pct@rev.desc.d9.sspro.natural.grid2-d3.mean-requests\endcsname{385.2}
\expandafter\def\csname ppt@rev.desc.d9.sspro.natural.grid2-d3.mean-requests\endcsname{$+$385.2}
\expandafter\def\csname val@rev.desc.d9.sspro.natural.grid2-d3.n-jobs\endcsname{432}
\expandafter\def\csname cnt@rev.desc.d9.sspro.natural.grid2-d3.n-jobs\endcsname{432}
\expandafter\def\csname pct@rev.desc.d9.sspro.natural.grid2-d3.n-jobs\endcsname{43200.0}
\expandafter\def\csname ppt@rev.desc.d9.sspro.natural.grid2-d3.n-jobs\endcsname{$+$43200.0}
\expandafter\def\csname val@rev.desc.d9.sspro.natural.grid2-d3.n-requests-match-scores\endcsname{1}
\expandafter\def\csname cnt@rev.desc.d9.sspro.natural.grid2-d3.n-requests-match-scores\endcsname{1}
\expandafter\def\csname pct@rev.desc.d9.sspro.natural.grid2-d3.n-requests-match-scores\endcsname{100.0}
\expandafter\def\csname ppt@rev.desc.d9.sspro.natural.grid2-d3.n-requests-match-scores\endcsname{$+$100.0}
\expandafter\def\csname val@rev.desc.d9.sspro.natural.grid2-d3.opus.mean-mpix\endcsname{3.538}
\expandafter\def\csname cnt@rev.desc.d9.sspro.natural.grid2-d3.opus.mean-mpix\endcsname{4}
\expandafter\def\csname pct@rev.desc.d9.sspro.natural.grid2-d3.opus.mean-mpix\endcsname{353.8}
\expandafter\def\csname ppt@rev.desc.d9.sspro.natural.grid2-d3.opus.mean-mpix\endcsname{$+$353.8}
\expandafter\def\csname val@rev.desc.d9.sspro.natural.grid2-d3.opus.mean-pixels\endcsname{3538388.994}
\expandafter\def\csname cnt@rev.desc.d9.sspro.natural.grid2-d3.opus.mean-pixels\endcsname{3{,}538{,}389}
\expandafter\def\csname pct@rev.desc.d9.sspro.natural.grid2-d3.opus.mean-pixels\endcsname{353838899.4}
\expandafter\def\csname ppt@rev.desc.d9.sspro.natural.grid2-d3.opus.mean-pixels\endcsname{$+$353838899.4}
\expandafter\def\csname val@rev.desc.d9.sspro.natural.grid2-d3.opus.mean-reader-mpix\endcsname{0.474}
\expandafter\def\csname cnt@rev.desc.d9.sspro.natural.grid2-d3.opus.mean-reader-mpix\endcsname{0}
\expandafter\def\csname pct@rev.desc.d9.sspro.natural.grid2-d3.opus.mean-reader-mpix\endcsname{47.4}
\expandafter\def\csname ppt@rev.desc.d9.sspro.natural.grid2-d3.opus.mean-reader-mpix\endcsname{$+$47.4}
\expandafter\def\csname val@rev.desc.d9.sspro.natural.grid2-d3.opus.mean-requests\endcsname{3.756}
\expandafter\def\csname cnt@rev.desc.d9.sspro.natural.grid2-d3.opus.mean-requests\endcsname{4}
\expandafter\def\csname pct@rev.desc.d9.sspro.natural.grid2-d3.opus.mean-requests\endcsname{375.6}
\expandafter\def\csname ppt@rev.desc.d9.sspro.natural.grid2-d3.opus.mean-requests\endcsname{$+$375.6}
\expandafter\def\csname val@rev.desc.d9.sspro.natural.grid2-d3.opus.n-jobs\endcsname{176}
\expandafter\def\csname cnt@rev.desc.d9.sspro.natural.grid2-d3.opus.n-jobs\endcsname{176}
\expandafter\def\csname pct@rev.desc.d9.sspro.natural.grid2-d3.opus.n-jobs\endcsname{17600.0}
\expandafter\def\csname ppt@rev.desc.d9.sspro.natural.grid2-d3.opus.n-jobs\endcsname{$+$17600.0}
\expandafter\def\csname val@rev.desc.d9.sspro.natural.grid2-d3.opus.n-requests-match-scores\endcsname{1}
\expandafter\def\csname cnt@rev.desc.d9.sspro.natural.grid2-d3.opus.n-requests-match-scores\endcsname{1}
\expandafter\def\csname pct@rev.desc.d9.sspro.natural.grid2-d3.opus.n-requests-match-scores\endcsname{100.0}
\expandafter\def\csname ppt@rev.desc.d9.sspro.natural.grid2-d3.opus.n-requests-match-scores\endcsname{$+$100.0}
\expandafter\def\csname val@rev.desc.d9.sspro.natural.grid2-d3.opus.ratio-to-direct\endcsname{6.056}
\expandafter\def\csname cnt@rev.desc.d9.sspro.natural.grid2-d3.opus.ratio-to-direct\endcsname{6}
\expandafter\def\csname pct@rev.desc.d9.sspro.natural.grid2-d3.opus.ratio-to-direct\endcsname{605.6}
\expandafter\def\csname ppt@rev.desc.d9.sspro.natural.grid2-d3.opus.ratio-to-direct\endcsname{$+$605.6}
\expandafter\def\csname val@rev.desc.d9.sspro.natural.grid2-d3.ratio-to-direct\endcsname{6.378}
\expandafter\def\csname cnt@rev.desc.d9.sspro.natural.grid2-d3.ratio-to-direct\endcsname{6}
\expandafter\def\csname pct@rev.desc.d9.sspro.natural.grid2-d3.ratio-to-direct\endcsname{637.8}
\expandafter\def\csname ppt@rev.desc.d9.sspro.natural.grid2-d3.ratio-to-direct\endcsname{$+$637.8}
\expandafter\def\csname val@rev.desc.d9.sspro.natural.grid2-d3.sonnet.mean-mpix\endcsname{3.747}
\expandafter\def\csname cnt@rev.desc.d9.sspro.natural.grid2-d3.sonnet.mean-mpix\endcsname{4}
\expandafter\def\csname pct@rev.desc.d9.sspro.natural.grid2-d3.sonnet.mean-mpix\endcsname{374.7}
\expandafter\def\csname ppt@rev.desc.d9.sspro.natural.grid2-d3.sonnet.mean-mpix\endcsname{$+$374.7}
\expandafter\def\csname val@rev.desc.d9.sspro.natural.grid2-d3.sonnet.mean-pixels\endcsname{3746776.011}
\expandafter\def\csname cnt@rev.desc.d9.sspro.natural.grid2-d3.sonnet.mean-pixels\endcsname{3{,}746{,}776}
\expandafter\def\csname pct@rev.desc.d9.sspro.natural.grid2-d3.sonnet.mean-pixels\endcsname{374677601.1}
\expandafter\def\csname ppt@rev.desc.d9.sspro.natural.grid2-d3.sonnet.mean-pixels\endcsname{$+$374677601.1}
\expandafter\def\csname val@rev.desc.d9.sspro.natural.grid2-d3.sonnet.mean-reader-mpix\endcsname{0.440}
\expandafter\def\csname cnt@rev.desc.d9.sspro.natural.grid2-d3.sonnet.mean-reader-mpix\endcsname{0}
\expandafter\def\csname pct@rev.desc.d9.sspro.natural.grid2-d3.sonnet.mean-reader-mpix\endcsname{44.0}
\expandafter\def\csname ppt@rev.desc.d9.sspro.natural.grid2-d3.sonnet.mean-reader-mpix\endcsname{$+$44.0}
\expandafter\def\csname val@rev.desc.d9.sspro.natural.grid2-d3.sonnet.mean-requests\endcsname{3.886}
\expandafter\def\csname cnt@rev.desc.d9.sspro.natural.grid2-d3.sonnet.mean-requests\endcsname{4}
\expandafter\def\csname pct@rev.desc.d9.sspro.natural.grid2-d3.sonnet.mean-requests\endcsname{388.6}
\expandafter\def\csname ppt@rev.desc.d9.sspro.natural.grid2-d3.sonnet.mean-requests\endcsname{$+$388.6}
\expandafter\def\csname val@rev.desc.d9.sspro.natural.grid2-d3.sonnet.n-jobs\endcsname{176}
\expandafter\def\csname cnt@rev.desc.d9.sspro.natural.grid2-d3.sonnet.n-jobs\endcsname{176}
\expandafter\def\csname pct@rev.desc.d9.sspro.natural.grid2-d3.sonnet.n-jobs\endcsname{17600.0}
\expandafter\def\csname ppt@rev.desc.d9.sspro.natural.grid2-d3.sonnet.n-jobs\endcsname{$+$17600.0}
\expandafter\def\csname val@rev.desc.d9.sspro.natural.grid2-d3.sonnet.n-requests-match-scores\endcsname{1}
\expandafter\def\csname cnt@rev.desc.d9.sspro.natural.grid2-d3.sonnet.n-requests-match-scores\endcsname{1}
\expandafter\def\csname pct@rev.desc.d9.sspro.natural.grid2-d3.sonnet.n-requests-match-scores\endcsname{100.0}
\expandafter\def\csname ppt@rev.desc.d9.sspro.natural.grid2-d3.sonnet.n-requests-match-scores\endcsname{$+$100.0}
\expandafter\def\csname val@rev.desc.d9.sspro.natural.grid2-d3.sonnet.ratio-to-direct\endcsname{6.413}
\expandafter\def\csname cnt@rev.desc.d9.sspro.natural.grid2-d3.sonnet.ratio-to-direct\endcsname{6}
\expandafter\def\csname pct@rev.desc.d9.sspro.natural.grid2-d3.sonnet.ratio-to-direct\endcsname{641.3}
\expandafter\def\csname ppt@rev.desc.d9.sspro.natural.grid2-d3.sonnet.ratio-to-direct\endcsname{$+$641.3}
\expandafter\def\csname val@rev.desc.d9.sspro.natural.zoomclick-d3.haiku.mean-mpix\endcsname{1.652}
\expandafter\def\csname cnt@rev.desc.d9.sspro.natural.zoomclick-d3.haiku.mean-mpix\endcsname{2}
\expandafter\def\csname pct@rev.desc.d9.sspro.natural.zoomclick-d3.haiku.mean-mpix\endcsname{165.2}
\expandafter\def\csname ppt@rev.desc.d9.sspro.natural.zoomclick-d3.haiku.mean-mpix\endcsname{$+$165.2}
\expandafter\def\csname val@rev.desc.d9.sspro.natural.zoomclick-d3.haiku.mean-pixels\endcsname{1652292.363}
\expandafter\def\csname cnt@rev.desc.d9.sspro.natural.zoomclick-d3.haiku.mean-pixels\endcsname{1{,}652{,}292}
\expandafter\def\csname pct@rev.desc.d9.sspro.natural.zoomclick-d3.haiku.mean-pixels\endcsname{165229236.2}
\expandafter\def\csname ppt@rev.desc.d9.sspro.natural.zoomclick-d3.haiku.mean-pixels\endcsname{$+$165229236.2}
\expandafter\def\csname val@rev.desc.d9.sspro.natural.zoomclick-d3.haiku.mean-reader-mpix\endcsname{0.514}
\expandafter\def\csname cnt@rev.desc.d9.sspro.natural.zoomclick-d3.haiku.mean-reader-mpix\endcsname{1}
\expandafter\def\csname pct@rev.desc.d9.sspro.natural.zoomclick-d3.haiku.mean-reader-mpix\endcsname{51.4}
\expandafter\def\csname ppt@rev.desc.d9.sspro.natural.zoomclick-d3.haiku.mean-reader-mpix\endcsname{$+$51.4}
\expandafter\def\csname val@rev.desc.d9.sspro.natural.zoomclick-d3.haiku.mean-requests\endcsname{2.913}
\expandafter\def\csname cnt@rev.desc.d9.sspro.natural.zoomclick-d3.haiku.mean-requests\endcsname{3}
\expandafter\def\csname pct@rev.desc.d9.sspro.natural.zoomclick-d3.haiku.mean-requests\endcsname{291.2}
\expandafter\def\csname ppt@rev.desc.d9.sspro.natural.zoomclick-d3.haiku.mean-requests\endcsname{$+$291.2}
\expandafter\def\csname val@rev.desc.d9.sspro.natural.zoomclick-d3.haiku.n-jobs\endcsname{80}
\expandafter\def\csname cnt@rev.desc.d9.sspro.natural.zoomclick-d3.haiku.n-jobs\endcsname{80}
\expandafter\def\csname pct@rev.desc.d9.sspro.natural.zoomclick-d3.haiku.n-jobs\endcsname{8000.0}
\expandafter\def\csname ppt@rev.desc.d9.sspro.natural.zoomclick-d3.haiku.n-jobs\endcsname{$+$8000.0}
\expandafter\def\csname val@rev.desc.d9.sspro.natural.zoomclick-d3.haiku.n-requests-match-scores\endcsname{1}
\expandafter\def\csname cnt@rev.desc.d9.sspro.natural.zoomclick-d3.haiku.n-requests-match-scores\endcsname{1}
\expandafter\def\csname pct@rev.desc.d9.sspro.natural.zoomclick-d3.haiku.n-requests-match-scores\endcsname{100.0}
\expandafter\def\csname ppt@rev.desc.d9.sspro.natural.zoomclick-d3.haiku.n-requests-match-scores\endcsname{$+$100.0}
\expandafter\def\csname val@rev.desc.d9.sspro.natural.zoomclick-d3.haiku.ratio-to-direct\endcsname{2.828}
\expandafter\def\csname cnt@rev.desc.d9.sspro.natural.zoomclick-d3.haiku.ratio-to-direct\endcsname{3}
\expandafter\def\csname pct@rev.desc.d9.sspro.natural.zoomclick-d3.haiku.ratio-to-direct\endcsname{282.8}
\expandafter\def\csname ppt@rev.desc.d9.sspro.natural.zoomclick-d3.haiku.ratio-to-direct\endcsname{$+$282.8}
\expandafter\def\csname val@rev.desc.d9.sspro.natural.zoomclick-d3.mean-mpix\endcsname{1.659}
\expandafter\def\csname cnt@rev.desc.d9.sspro.natural.zoomclick-d3.mean-mpix\endcsname{2}
\expandafter\def\csname pct@rev.desc.d9.sspro.natural.zoomclick-d3.mean-mpix\endcsname{165.9}
\expandafter\def\csname ppt@rev.desc.d9.sspro.natural.zoomclick-d3.mean-mpix\endcsname{$+$165.9}
\expandafter\def\csname val@rev.desc.d9.sspro.natural.zoomclick-d3.mean-pixels\endcsname{1659244.588}
\expandafter\def\csname cnt@rev.desc.d9.sspro.natural.zoomclick-d3.mean-pixels\endcsname{1{,}659{,}245}
\expandafter\def\csname pct@rev.desc.d9.sspro.natural.zoomclick-d3.mean-pixels\endcsname{165924458.8}
\expandafter\def\csname ppt@rev.desc.d9.sspro.natural.zoomclick-d3.mean-pixels\endcsname{$+$165924458.8}
\expandafter\def\csname val@rev.desc.d9.sspro.natural.zoomclick-d3.mean-reader-mpix\endcsname{0.496}
\expandafter\def\csname cnt@rev.desc.d9.sspro.natural.zoomclick-d3.mean-reader-mpix\endcsname{0}
\expandafter\def\csname pct@rev.desc.d9.sspro.natural.zoomclick-d3.mean-reader-mpix\endcsname{49.6}
\expandafter\def\csname ppt@rev.desc.d9.sspro.natural.zoomclick-d3.mean-reader-mpix\endcsname{$+$49.6}
\expandafter\def\csname val@rev.desc.d9.sspro.natural.zoomclick-d3.mean-requests\endcsname{2.979}
\expandafter\def\csname cnt@rev.desc.d9.sspro.natural.zoomclick-d3.mean-requests\endcsname{3}
\expandafter\def\csname pct@rev.desc.d9.sspro.natural.zoomclick-d3.mean-requests\endcsname{297.9}
\expandafter\def\csname ppt@rev.desc.d9.sspro.natural.zoomclick-d3.mean-requests\endcsname{$+$297.9}
\expandafter\def\csname val@rev.desc.d9.sspro.natural.zoomclick-d3.n-jobs\endcsname{432}
\expandafter\def\csname cnt@rev.desc.d9.sspro.natural.zoomclick-d3.n-jobs\endcsname{432}
\expandafter\def\csname pct@rev.desc.d9.sspro.natural.zoomclick-d3.n-jobs\endcsname{43200.0}
\expandafter\def\csname ppt@rev.desc.d9.sspro.natural.zoomclick-d3.n-jobs\endcsname{$+$43200.0}
\expandafter\def\csname val@rev.desc.d9.sspro.natural.zoomclick-d3.n-requests-match-scores\endcsname{1}
\expandafter\def\csname cnt@rev.desc.d9.sspro.natural.zoomclick-d3.n-requests-match-scores\endcsname{1}
\expandafter\def\csname pct@rev.desc.d9.sspro.natural.zoomclick-d3.n-requests-match-scores\endcsname{100.0}
\expandafter\def\csname ppt@rev.desc.d9.sspro.natural.zoomclick-d3.n-requests-match-scores\endcsname{$+$100.0}
\expandafter\def\csname val@rev.desc.d9.sspro.natural.zoomclick-d3.opus.mean-mpix\endcsname{1.662}
\expandafter\def\csname cnt@rev.desc.d9.sspro.natural.zoomclick-d3.opus.mean-mpix\endcsname{2}
\expandafter\def\csname pct@rev.desc.d9.sspro.natural.zoomclick-d3.opus.mean-mpix\endcsname{166.2}
\expandafter\def\csname ppt@rev.desc.d9.sspro.natural.zoomclick-d3.opus.mean-mpix\endcsname{$+$166.2}
\expandafter\def\csname val@rev.desc.d9.sspro.natural.zoomclick-d3.opus.mean-pixels\endcsname{1662013.494}
\expandafter\def\csname cnt@rev.desc.d9.sspro.natural.zoomclick-d3.opus.mean-pixels\endcsname{1{,}662{,}013}
\expandafter\def\csname pct@rev.desc.d9.sspro.natural.zoomclick-d3.opus.mean-pixels\endcsname{166201349.4}
\expandafter\def\csname ppt@rev.desc.d9.sspro.natural.zoomclick-d3.opus.mean-pixels\endcsname{$+$166201349.4}
\expandafter\def\csname val@rev.desc.d9.sspro.natural.zoomclick-d3.opus.mean-reader-mpix\endcsname{0.492}
\expandafter\def\csname cnt@rev.desc.d9.sspro.natural.zoomclick-d3.opus.mean-reader-mpix\endcsname{0}
\expandafter\def\csname pct@rev.desc.d9.sspro.natural.zoomclick-d3.opus.mean-reader-mpix\endcsname{49.2}
\expandafter\def\csname ppt@rev.desc.d9.sspro.natural.zoomclick-d3.opus.mean-reader-mpix\endcsname{$+$49.2}
\expandafter\def\csname val@rev.desc.d9.sspro.natural.zoomclick-d3.opus.mean-requests\endcsname{3.000}
\expandafter\def\csname cnt@rev.desc.d9.sspro.natural.zoomclick-d3.opus.mean-requests\endcsname{3}
\expandafter\def\csname pct@rev.desc.d9.sspro.natural.zoomclick-d3.opus.mean-requests\endcsname{300.0}
\expandafter\def\csname ppt@rev.desc.d9.sspro.natural.zoomclick-d3.opus.mean-requests\endcsname{$+$300.0}
\expandafter\def\csname val@rev.desc.d9.sspro.natural.zoomclick-d3.opus.n-jobs\endcsname{176}
\expandafter\def\csname cnt@rev.desc.d9.sspro.natural.zoomclick-d3.opus.n-jobs\endcsname{176}
\expandafter\def\csname pct@rev.desc.d9.sspro.natural.zoomclick-d3.opus.n-jobs\endcsname{17600.0}
\expandafter\def\csname ppt@rev.desc.d9.sspro.natural.zoomclick-d3.opus.n-jobs\endcsname{$+$17600.0}
\expandafter\def\csname val@rev.desc.d9.sspro.natural.zoomclick-d3.opus.n-requests-match-scores\endcsname{1}
\expandafter\def\csname cnt@rev.desc.d9.sspro.natural.zoomclick-d3.opus.n-requests-match-scores\endcsname{1}
\expandafter\def\csname pct@rev.desc.d9.sspro.natural.zoomclick-d3.opus.n-requests-match-scores\endcsname{100.0}
\expandafter\def\csname ppt@rev.desc.d9.sspro.natural.zoomclick-d3.opus.n-requests-match-scores\endcsname{$+$100.0}
\expandafter\def\csname val@rev.desc.d9.sspro.natural.zoomclick-d3.opus.ratio-to-direct\endcsname{2.845}
\expandafter\def\csname cnt@rev.desc.d9.sspro.natural.zoomclick-d3.opus.ratio-to-direct\endcsname{3}
\expandafter\def\csname pct@rev.desc.d9.sspro.natural.zoomclick-d3.opus.ratio-to-direct\endcsname{284.5}
\expandafter\def\csname ppt@rev.desc.d9.sspro.natural.zoomclick-d3.opus.ratio-to-direct\endcsname{$+$284.5}
\expandafter\def\csname val@rev.desc.d9.sspro.natural.zoomclick-d3.ratio-to-direct\endcsname{2.840}
\expandafter\def\csname cnt@rev.desc.d9.sspro.natural.zoomclick-d3.ratio-to-direct\endcsname{3}
\expandafter\def\csname pct@rev.desc.d9.sspro.natural.zoomclick-d3.ratio-to-direct\endcsname{284.0}
\expandafter\def\csname ppt@rev.desc.d9.sspro.natural.zoomclick-d3.ratio-to-direct\endcsname{$+$284.0}
\expandafter\def\csname val@rev.desc.d9.sspro.natural.zoomclick-d3.sonnet.mean-mpix\endcsname{1.660}
\expandafter\def\csname cnt@rev.desc.d9.sspro.natural.zoomclick-d3.sonnet.mean-mpix\endcsname{2}
\expandafter\def\csname pct@rev.desc.d9.sspro.natural.zoomclick-d3.sonnet.mean-mpix\endcsname{166.0}
\expandafter\def\csname ppt@rev.desc.d9.sspro.natural.zoomclick-d3.sonnet.mean-mpix\endcsname{$+$166.0}
\expandafter\def\csname val@rev.desc.d9.sspro.natural.zoomclick-d3.sonnet.mean-pixels\endcsname{1659635.784}
\expandafter\def\csname cnt@rev.desc.d9.sspro.natural.zoomclick-d3.sonnet.mean-pixels\endcsname{1{,}659{,}636}
\expandafter\def\csname pct@rev.desc.d9.sspro.natural.zoomclick-d3.sonnet.mean-pixels\endcsname{165963578.4}
\expandafter\def\csname ppt@rev.desc.d9.sspro.natural.zoomclick-d3.sonnet.mean-pixels\endcsname{$+$165963578.4}
\expandafter\def\csname val@rev.desc.d9.sspro.natural.zoomclick-d3.sonnet.mean-reader-mpix\endcsname{0.491}
\expandafter\def\csname cnt@rev.desc.d9.sspro.natural.zoomclick-d3.sonnet.mean-reader-mpix\endcsname{0}
\expandafter\def\csname pct@rev.desc.d9.sspro.natural.zoomclick-d3.sonnet.mean-reader-mpix\endcsname{49.1}
\expandafter\def\csname ppt@rev.desc.d9.sspro.natural.zoomclick-d3.sonnet.mean-reader-mpix\endcsname{$+$49.1}
\expandafter\def\csname val@rev.desc.d9.sspro.natural.zoomclick-d3.sonnet.mean-requests\endcsname{2.989}
\expandafter\def\csname cnt@rev.desc.d9.sspro.natural.zoomclick-d3.sonnet.mean-requests\endcsname{3}
\expandafter\def\csname pct@rev.desc.d9.sspro.natural.zoomclick-d3.sonnet.mean-requests\endcsname{298.9}
\expandafter\def\csname ppt@rev.desc.d9.sspro.natural.zoomclick-d3.sonnet.mean-requests\endcsname{$+$298.9}
\expandafter\def\csname val@rev.desc.d9.sspro.natural.zoomclick-d3.sonnet.n-jobs\endcsname{176}
\expandafter\def\csname cnt@rev.desc.d9.sspro.natural.zoomclick-d3.sonnet.n-jobs\endcsname{176}
\expandafter\def\csname pct@rev.desc.d9.sspro.natural.zoomclick-d3.sonnet.n-jobs\endcsname{17600.0}
\expandafter\def\csname ppt@rev.desc.d9.sspro.natural.zoomclick-d3.sonnet.n-jobs\endcsname{$+$17600.0}
\expandafter\def\csname val@rev.desc.d9.sspro.natural.zoomclick-d3.sonnet.n-requests-match-scores\endcsname{1}
\expandafter\def\csname cnt@rev.desc.d9.sspro.natural.zoomclick-d3.sonnet.n-requests-match-scores\endcsname{1}
\expandafter\def\csname pct@rev.desc.d9.sspro.natural.zoomclick-d3.sonnet.n-requests-match-scores\endcsname{100.0}
\expandafter\def\csname ppt@rev.desc.d9.sspro.natural.zoomclick-d3.sonnet.n-requests-match-scores\endcsname{$+$100.0}
\expandafter\def\csname val@rev.desc.d9.sspro.natural.zoomclick-d3.sonnet.ratio-to-direct\endcsname{2.841}
\expandafter\def\csname cnt@rev.desc.d9.sspro.natural.zoomclick-d3.sonnet.ratio-to-direct\endcsname{3}
\expandafter\def\csname pct@rev.desc.d9.sspro.natural.zoomclick-d3.sonnet.ratio-to-direct\endcsname{284.1}
\expandafter\def\csname ppt@rev.desc.d9.sspro.natural.zoomclick-d3.sonnet.ratio-to-direct\endcsname{$+$284.1}
\expandafter\def\csname val@rev.desc.d9.sspro.replay.haiku.mean-mpix\endcsname{0.512}
\expandafter\def\csname cnt@rev.desc.d9.sspro.replay.haiku.mean-mpix\endcsname{1}
\expandafter\def\csname pct@rev.desc.d9.sspro.replay.haiku.mean-mpix\endcsname{51.2}
\expandafter\def\csname ppt@rev.desc.d9.sspro.replay.haiku.mean-mpix\endcsname{$+$51.2}
\expandafter\def\csname val@rev.desc.d9.sspro.replay.haiku.mean-pixels\endcsname{511777.125}
\expandafter\def\csname cnt@rev.desc.d9.sspro.replay.haiku.mean-pixels\endcsname{511{,}777}
\expandafter\def\csname pct@rev.desc.d9.sspro.replay.haiku.mean-pixels\endcsname{51177712.5}
\expandafter\def\csname ppt@rev.desc.d9.sspro.replay.haiku.mean-pixels\endcsname{$+$51177712.5}
\expandafter\def\csname val@rev.desc.d9.sspro.replay.haiku.mean-reader-mpix\endcsname{0.512}
\expandafter\def\csname cnt@rev.desc.d9.sspro.replay.haiku.mean-reader-mpix\endcsname{1}
\expandafter\def\csname pct@rev.desc.d9.sspro.replay.haiku.mean-reader-mpix\endcsname{51.2}
\expandafter\def\csname ppt@rev.desc.d9.sspro.replay.haiku.mean-reader-mpix\endcsname{$+$51.2}
\expandafter\def\csname val@rev.desc.d9.sspro.replay.haiku.mean-requests\endcsname{1.000}
\expandafter\def\csname cnt@rev.desc.d9.sspro.replay.haiku.mean-requests\endcsname{1}
\expandafter\def\csname pct@rev.desc.d9.sspro.replay.haiku.mean-requests\endcsname{100.0}
\expandafter\def\csname ppt@rev.desc.d9.sspro.replay.haiku.mean-requests\endcsname{$+$100.0}
\expandafter\def\csname val@rev.desc.d9.sspro.replay.haiku.n-jobs\endcsname{240}
\expandafter\def\csname cnt@rev.desc.d9.sspro.replay.haiku.n-jobs\endcsname{240}
\expandafter\def\csname pct@rev.desc.d9.sspro.replay.haiku.n-jobs\endcsname{24000.0}
\expandafter\def\csname ppt@rev.desc.d9.sspro.replay.haiku.n-jobs\endcsname{$+$24000.0}
\expandafter\def\csname val@rev.desc.d9.sspro.replay.haiku.n-requests-match-scores\endcsname{1}
\expandafter\def\csname cnt@rev.desc.d9.sspro.replay.haiku.n-requests-match-scores\endcsname{1}
\expandafter\def\csname pct@rev.desc.d9.sspro.replay.haiku.n-requests-match-scores\endcsname{100.0}
\expandafter\def\csname ppt@rev.desc.d9.sspro.replay.haiku.n-requests-match-scores\endcsname{$+$100.0}
\expandafter\def\csname val@rev.desc.d9.sspro.replay.haiku.ratio-to-direct\endcsname{0.876}
\expandafter\def\csname cnt@rev.desc.d9.sspro.replay.haiku.ratio-to-direct\endcsname{1}
\expandafter\def\csname pct@rev.desc.d9.sspro.replay.haiku.ratio-to-direct\endcsname{87.6}
\expandafter\def\csname ppt@rev.desc.d9.sspro.replay.haiku.ratio-to-direct\endcsname{$+$87.6}
\expandafter\def\csname val@rev.desc.d9.sspro.replay.mean-mpix\endcsname{0.531}
\expandafter\def\csname cnt@rev.desc.d9.sspro.replay.mean-mpix\endcsname{1}
\expandafter\def\csname pct@rev.desc.d9.sspro.replay.mean-mpix\endcsname{53.1}
\expandafter\def\csname ppt@rev.desc.d9.sspro.replay.mean-mpix\endcsname{$+$53.1}
\expandafter\def\csname val@rev.desc.d9.sspro.replay.mean-pixels\endcsname{531021.503}
\expandafter\def\csname cnt@rev.desc.d9.sspro.replay.mean-pixels\endcsname{531{,}022}
\expandafter\def\csname pct@rev.desc.d9.sspro.replay.mean-pixels\endcsname{53102150.3}
\expandafter\def\csname ppt@rev.desc.d9.sspro.replay.mean-pixels\endcsname{$+$53102150.3}
\expandafter\def\csname val@rev.desc.d9.sspro.replay.mean-reader-mpix\endcsname{0.531}
\expandafter\def\csname cnt@rev.desc.d9.sspro.replay.mean-reader-mpix\endcsname{1}
\expandafter\def\csname pct@rev.desc.d9.sspro.replay.mean-reader-mpix\endcsname{53.1}
\expandafter\def\csname ppt@rev.desc.d9.sspro.replay.mean-reader-mpix\endcsname{$+$53.1}
\expandafter\def\csname val@rev.desc.d9.sspro.replay.mean-requests\endcsname{1.000}
\expandafter\def\csname cnt@rev.desc.d9.sspro.replay.mean-requests\endcsname{1}
\expandafter\def\csname pct@rev.desc.d9.sspro.replay.mean-requests\endcsname{100.0}
\expandafter\def\csname ppt@rev.desc.d9.sspro.replay.mean-requests\endcsname{$+$100.0}
\expandafter\def\csname val@rev.desc.d9.sspro.replay.n-jobs\endcsname{1760}
\expandafter\def\csname cnt@rev.desc.d9.sspro.replay.n-jobs\endcsname{1{,}760}
\expandafter\def\csname pct@rev.desc.d9.sspro.replay.n-jobs\endcsname{176000.0}
\expandafter\def\csname ppt@rev.desc.d9.sspro.replay.n-jobs\endcsname{$+$176000.0}
\expandafter\def\csname val@rev.desc.d9.sspro.replay.n-requests-match-scores\endcsname{1}
\expandafter\def\csname cnt@rev.desc.d9.sspro.replay.n-requests-match-scores\endcsname{1}
\expandafter\def\csname pct@rev.desc.d9.sspro.replay.n-requests-match-scores\endcsname{100.0}
\expandafter\def\csname ppt@rev.desc.d9.sspro.replay.n-requests-match-scores\endcsname{$+$100.0}
\expandafter\def\csname val@rev.desc.d9.sspro.replay.opus.mean-mpix\endcsname{0.528}
\expandafter\def\csname cnt@rev.desc.d9.sspro.replay.opus.mean-mpix\endcsname{1}
\expandafter\def\csname pct@rev.desc.d9.sspro.replay.opus.mean-mpix\endcsname{52.8}
\expandafter\def\csname ppt@rev.desc.d9.sspro.replay.opus.mean-mpix\endcsname{$+$52.8}
\expandafter\def\csname val@rev.desc.d9.sspro.replay.opus.mean-pixels\endcsname{527789.361}
\expandafter\def\csname cnt@rev.desc.d9.sspro.replay.opus.mean-pixels\endcsname{527{,}789}
\expandafter\def\csname pct@rev.desc.d9.sspro.replay.opus.mean-pixels\endcsname{52778936.1}
\expandafter\def\csname ppt@rev.desc.d9.sspro.replay.opus.mean-pixels\endcsname{$+$52778936.1}
\expandafter\def\csname val@rev.desc.d9.sspro.replay.opus.mean-reader-mpix\endcsname{0.528}
\expandafter\def\csname cnt@rev.desc.d9.sspro.replay.opus.mean-reader-mpix\endcsname{1}
\expandafter\def\csname pct@rev.desc.d9.sspro.replay.opus.mean-reader-mpix\endcsname{52.8}
\expandafter\def\csname ppt@rev.desc.d9.sspro.replay.opus.mean-reader-mpix\endcsname{$+$52.8}
\expandafter\def\csname val@rev.desc.d9.sspro.replay.opus.mean-requests\endcsname{1.000}
\expandafter\def\csname cnt@rev.desc.d9.sspro.replay.opus.mean-requests\endcsname{1}
\expandafter\def\csname pct@rev.desc.d9.sspro.replay.opus.mean-requests\endcsname{100.0}
\expandafter\def\csname ppt@rev.desc.d9.sspro.replay.opus.mean-requests\endcsname{$+$100.0}
\expandafter\def\csname val@rev.desc.d9.sspro.replay.opus.n-jobs\endcsname{880}
\expandafter\def\csname cnt@rev.desc.d9.sspro.replay.opus.n-jobs\endcsname{880}
\expandafter\def\csname pct@rev.desc.d9.sspro.replay.opus.n-jobs\endcsname{88000.0}
\expandafter\def\csname ppt@rev.desc.d9.sspro.replay.opus.n-jobs\endcsname{$+$88000.0}
\expandafter\def\csname val@rev.desc.d9.sspro.replay.opus.n-requests-match-scores\endcsname{1}
\expandafter\def\csname cnt@rev.desc.d9.sspro.replay.opus.n-requests-match-scores\endcsname{1}
\expandafter\def\csname pct@rev.desc.d9.sspro.replay.opus.n-requests-match-scores\endcsname{100.0}
\expandafter\def\csname ppt@rev.desc.d9.sspro.replay.opus.n-requests-match-scores\endcsname{$+$100.0}
\expandafter\def\csname val@rev.desc.d9.sspro.replay.opus.ratio-to-direct\endcsname{0.903}
\expandafter\def\csname cnt@rev.desc.d9.sspro.replay.opus.ratio-to-direct\endcsname{1}
\expandafter\def\csname pct@rev.desc.d9.sspro.replay.opus.ratio-to-direct\endcsname{90.3}
\expandafter\def\csname ppt@rev.desc.d9.sspro.replay.opus.ratio-to-direct\endcsname{$+$90.3}
\expandafter\def\csname val@rev.desc.d9.sspro.replay.ratio-to-direct\endcsname{0.909}
\expandafter\def\csname cnt@rev.desc.d9.sspro.replay.ratio-to-direct\endcsname{1}
\expandafter\def\csname pct@rev.desc.d9.sspro.replay.ratio-to-direct\endcsname{90.9}
\expandafter\def\csname ppt@rev.desc.d9.sspro.replay.ratio-to-direct\endcsname{$+$90.9}
\expandafter\def\csname val@rev.desc.d9.sspro.replay.sonnet.mean-mpix\endcsname{0.543}
\expandafter\def\csname cnt@rev.desc.d9.sspro.replay.sonnet.mean-mpix\endcsname{1}
\expandafter\def\csname pct@rev.desc.d9.sspro.replay.sonnet.mean-mpix\endcsname{54.3}
\expandafter\def\csname ppt@rev.desc.d9.sspro.replay.sonnet.mean-mpix\endcsname{$+$54.3}
\expandafter\def\csname val@rev.desc.d9.sspro.replay.sonnet.mean-pixels\endcsname{542682.341}
\expandafter\def\csname cnt@rev.desc.d9.sspro.replay.sonnet.mean-pixels\endcsname{542{,}682}
\expandafter\def\csname pct@rev.desc.d9.sspro.replay.sonnet.mean-pixels\endcsname{54268234.1}
\expandafter\def\csname ppt@rev.desc.d9.sspro.replay.sonnet.mean-pixels\endcsname{$+$54268234.1}
\expandafter\def\csname val@rev.desc.d9.sspro.replay.sonnet.mean-reader-mpix\endcsname{0.543}
\expandafter\def\csname cnt@rev.desc.d9.sspro.replay.sonnet.mean-reader-mpix\endcsname{1}
\expandafter\def\csname pct@rev.desc.d9.sspro.replay.sonnet.mean-reader-mpix\endcsname{54.3}
\expandafter\def\csname ppt@rev.desc.d9.sspro.replay.sonnet.mean-reader-mpix\endcsname{$+$54.3}
\expandafter\def\csname val@rev.desc.d9.sspro.replay.sonnet.mean-requests\endcsname{1.000}
\expandafter\def\csname cnt@rev.desc.d9.sspro.replay.sonnet.mean-requests\endcsname{1}
\expandafter\def\csname pct@rev.desc.d9.sspro.replay.sonnet.mean-requests\endcsname{100.0}
\expandafter\def\csname ppt@rev.desc.d9.sspro.replay.sonnet.mean-requests\endcsname{$+$100.0}
\expandafter\def\csname val@rev.desc.d9.sspro.replay.sonnet.n-jobs\endcsname{640}
\expandafter\def\csname cnt@rev.desc.d9.sspro.replay.sonnet.n-jobs\endcsname{640}
\expandafter\def\csname pct@rev.desc.d9.sspro.replay.sonnet.n-jobs\endcsname{64000.0}
\expandafter\def\csname ppt@rev.desc.d9.sspro.replay.sonnet.n-jobs\endcsname{$+$64000.0}
\expandafter\def\csname val@rev.desc.d9.sspro.replay.sonnet.n-requests-match-scores\endcsname{1}
\expandafter\def\csname cnt@rev.desc.d9.sspro.replay.sonnet.n-requests-match-scores\endcsname{1}
\expandafter\def\csname pct@rev.desc.d9.sspro.replay.sonnet.n-requests-match-scores\endcsname{100.0}
\expandafter\def\csname ppt@rev.desc.d9.sspro.replay.sonnet.n-requests-match-scores\endcsname{$+$100.0}
\expandafter\def\csname val@rev.desc.d9.sspro.replay.sonnet.ratio-to-direct\endcsname{0.929}
\expandafter\def\csname cnt@rev.desc.d9.sspro.replay.sonnet.ratio-to-direct\endcsname{1}
\expandafter\def\csname pct@rev.desc.d9.sspro.replay.sonnet.ratio-to-direct\endcsname{92.9}
\expandafter\def\csname ppt@rev.desc.d9.sspro.replay.sonnet.ratio-to-direct\endcsname{$+$92.9}
\expandafter\def\csname val@rev.desc.d9.uiv.direct.mean-mpix\endcsname{0.587}
\expandafter\def\csname cnt@rev.desc.d9.uiv.direct.mean-mpix\endcsname{1}
\expandafter\def\csname pct@rev.desc.d9.uiv.direct.mean-mpix\endcsname{58.7}
\expandafter\def\csname ppt@rev.desc.d9.uiv.direct.mean-mpix\endcsname{$+$58.7}
\expandafter\def\csname val@rev.desc.d9.uiv.direct.mean-pixels\endcsname{586804.065}
\expandafter\def\csname cnt@rev.desc.d9.uiv.direct.mean-pixels\endcsname{586{,}804}
\expandafter\def\csname pct@rev.desc.d9.uiv.direct.mean-pixels\endcsname{58680406.5}
\expandafter\def\csname ppt@rev.desc.d9.uiv.direct.mean-pixels\endcsname{$+$58680406.5}
\expandafter\def\csname val@rev.desc.d9.uiv.direct.mean-reader-mpix\endcsname{0.000}
\expandafter\def\csname cnt@rev.desc.d9.uiv.direct.mean-reader-mpix\endcsname{0}
\expandafter\def\csname pct@rev.desc.d9.uiv.direct.mean-reader-mpix\endcsname{0.0}
\expandafter\def\csname ppt@rev.desc.d9.uiv.direct.mean-reader-mpix\endcsname{0.0}
\expandafter\def\csname val@rev.desc.d9.uiv.direct.mean-requests\endcsname{1.000}
\expandafter\def\csname cnt@rev.desc.d9.uiv.direct.mean-requests\endcsname{1}
\expandafter\def\csname pct@rev.desc.d9.uiv.direct.mean-requests\endcsname{100.0}
\expandafter\def\csname ppt@rev.desc.d9.uiv.direct.mean-requests\endcsname{$+$100.0}
\expandafter\def\csname val@rev.desc.d9.uiv.direct.n-jobs\endcsname{1656}
\expandafter\def\csname cnt@rev.desc.d9.uiv.direct.n-jobs\endcsname{1{,}656}
\expandafter\def\csname pct@rev.desc.d9.uiv.direct.n-jobs\endcsname{165600.0}
\expandafter\def\csname ppt@rev.desc.d9.uiv.direct.n-jobs\endcsname{$+$165600.0}
\expandafter\def\csname val@rev.desc.d9.uiv.direct.n-requests-match-scores\endcsname{1}
\expandafter\def\csname cnt@rev.desc.d9.uiv.direct.n-requests-match-scores\endcsname{1}
\expandafter\def\csname pct@rev.desc.d9.uiv.direct.n-requests-match-scores\endcsname{100.0}
\expandafter\def\csname ppt@rev.desc.d9.uiv.direct.n-requests-match-scores\endcsname{$+$100.0}
\expandafter\def\csname val@rev.desc.d9.uiv.direct.ratio-to-direct\endcsname{1.000}
\expandafter\def\csname cnt@rev.desc.d9.uiv.direct.ratio-to-direct\endcsname{1}
\expandafter\def\csname pct@rev.desc.d9.uiv.direct.ratio-to-direct\endcsname{100.0}
\expandafter\def\csname ppt@rev.desc.d9.uiv.direct.ratio-to-direct\endcsname{$+$100.0}
\expandafter\def\csname val@rev.desc.d9.uiv.natural.center-d1.haiku.mean-mpix\endcsname{1.117}
\expandafter\def\csname cnt@rev.desc.d9.uiv.natural.center-d1.haiku.mean-mpix\endcsname{1}
\expandafter\def\csname pct@rev.desc.d9.uiv.natural.center-d1.haiku.mean-mpix\endcsname{111.7}
\expandafter\def\csname ppt@rev.desc.d9.uiv.natural.center-d1.haiku.mean-mpix\endcsname{$+$111.7}
\expandafter\def\csname val@rev.desc.d9.uiv.natural.center-d1.haiku.mean-pixels\endcsname{1117093.731}
\expandafter\def\csname cnt@rev.desc.d9.uiv.natural.center-d1.haiku.mean-pixels\endcsname{1{,}117{,}094}
\expandafter\def\csname pct@rev.desc.d9.uiv.natural.center-d1.haiku.mean-pixels\endcsname{111709373.1}
\expandafter\def\csname ppt@rev.desc.d9.uiv.natural.center-d1.haiku.mean-pixels\endcsname{$+$111709373.1}
\expandafter\def\csname val@rev.desc.d9.uiv.natural.center-d1.haiku.mean-reader-mpix\endcsname{0.534}
\expandafter\def\csname cnt@rev.desc.d9.uiv.natural.center-d1.haiku.mean-reader-mpix\endcsname{1}
\expandafter\def\csname pct@rev.desc.d9.uiv.natural.center-d1.haiku.mean-reader-mpix\endcsname{53.4}
\expandafter\def\csname ppt@rev.desc.d9.uiv.natural.center-d1.haiku.mean-reader-mpix\endcsname{$+$53.4}
\expandafter\def\csname val@rev.desc.d9.uiv.natural.center-d1.haiku.mean-requests\endcsname{2.000}
\expandafter\def\csname cnt@rev.desc.d9.uiv.natural.center-d1.haiku.mean-requests\endcsname{2}
\expandafter\def\csname pct@rev.desc.d9.uiv.natural.center-d1.haiku.mean-requests\endcsname{200.0}
\expandafter\def\csname ppt@rev.desc.d9.uiv.natural.center-d1.haiku.mean-requests\endcsname{$+$200.0}
\expandafter\def\csname val@rev.desc.d9.uiv.natural.center-d1.haiku.n-jobs\endcsname{160}
\expandafter\def\csname cnt@rev.desc.d9.uiv.natural.center-d1.haiku.n-jobs\endcsname{160}
\expandafter\def\csname pct@rev.desc.d9.uiv.natural.center-d1.haiku.n-jobs\endcsname{16000.0}
\expandafter\def\csname ppt@rev.desc.d9.uiv.natural.center-d1.haiku.n-jobs\endcsname{$+$16000.0}
\expandafter\def\csname val@rev.desc.d9.uiv.natural.center-d1.haiku.n-requests-match-scores\endcsname{1}
\expandafter\def\csname cnt@rev.desc.d9.uiv.natural.center-d1.haiku.n-requests-match-scores\endcsname{1}
\expandafter\def\csname pct@rev.desc.d9.uiv.natural.center-d1.haiku.n-requests-match-scores\endcsname{100.0}
\expandafter\def\csname ppt@rev.desc.d9.uiv.natural.center-d1.haiku.n-requests-match-scores\endcsname{$+$100.0}
\expandafter\def\csname val@rev.desc.d9.uiv.natural.center-d1.haiku.ratio-to-direct\endcsname{1.904}
\expandafter\def\csname cnt@rev.desc.d9.uiv.natural.center-d1.haiku.ratio-to-direct\endcsname{2}
\expandafter\def\csname pct@rev.desc.d9.uiv.natural.center-d1.haiku.ratio-to-direct\endcsname{190.4}
\expandafter\def\csname ppt@rev.desc.d9.uiv.natural.center-d1.haiku.ratio-to-direct\endcsname{$+$190.4}
\expandafter\def\csname val@rev.desc.d9.uiv.natural.center-d1.mean-mpix\endcsname{1.125}
\expandafter\def\csname cnt@rev.desc.d9.uiv.natural.center-d1.mean-mpix\endcsname{1}
\expandafter\def\csname pct@rev.desc.d9.uiv.natural.center-d1.mean-mpix\endcsname{112.5}
\expandafter\def\csname ppt@rev.desc.d9.uiv.natural.center-d1.mean-mpix\endcsname{$+$112.5}
\expandafter\def\csname val@rev.desc.d9.uiv.natural.center-d1.mean-pixels\endcsname{1124504.650}
\expandafter\def\csname cnt@rev.desc.d9.uiv.natural.center-d1.mean-pixels\endcsname{1{,}124{,}505}
\expandafter\def\csname pct@rev.desc.d9.uiv.natural.center-d1.mean-pixels\endcsname{112450465.0}
\expandafter\def\csname ppt@rev.desc.d9.uiv.natural.center-d1.mean-pixels\endcsname{$+$112450465.0}
\expandafter\def\csname val@rev.desc.d9.uiv.natural.center-d1.mean-reader-mpix\endcsname{0.538}
\expandafter\def\csname cnt@rev.desc.d9.uiv.natural.center-d1.mean-reader-mpix\endcsname{1}
\expandafter\def\csname pct@rev.desc.d9.uiv.natural.center-d1.mean-reader-mpix\endcsname{53.8}
\expandafter\def\csname ppt@rev.desc.d9.uiv.natural.center-d1.mean-reader-mpix\endcsname{$+$53.8}
\expandafter\def\csname val@rev.desc.d9.uiv.natural.center-d1.mean-requests\endcsname{2.001}
\expandafter\def\csname cnt@rev.desc.d9.uiv.natural.center-d1.mean-requests\endcsname{2}
\expandafter\def\csname pct@rev.desc.d9.uiv.natural.center-d1.mean-requests\endcsname{200.1}
\expandafter\def\csname ppt@rev.desc.d9.uiv.natural.center-d1.mean-requests\endcsname{$+$200.1}
\expandafter\def\csname val@rev.desc.d9.uiv.natural.center-d1.n-jobs\endcsname{768}
\expandafter\def\csname cnt@rev.desc.d9.uiv.natural.center-d1.n-jobs\endcsname{768}
\expandafter\def\csname pct@rev.desc.d9.uiv.natural.center-d1.n-jobs\endcsname{76800.0}
\expandafter\def\csname ppt@rev.desc.d9.uiv.natural.center-d1.n-jobs\endcsname{$+$76800.0}
\expandafter\def\csname val@rev.desc.d9.uiv.natural.center-d1.n-requests-match-scores\endcsname{1}
\expandafter\def\csname cnt@rev.desc.d9.uiv.natural.center-d1.n-requests-match-scores\endcsname{1}
\expandafter\def\csname pct@rev.desc.d9.uiv.natural.center-d1.n-requests-match-scores\endcsname{100.0}
\expandafter\def\csname ppt@rev.desc.d9.uiv.natural.center-d1.n-requests-match-scores\endcsname{$+$100.0}
\expandafter\def\csname val@rev.desc.d9.uiv.natural.center-d1.opus.mean-mpix\endcsname{1.125}
\expandafter\def\csname cnt@rev.desc.d9.uiv.natural.center-d1.opus.mean-mpix\endcsname{1}
\expandafter\def\csname pct@rev.desc.d9.uiv.natural.center-d1.opus.mean-mpix\endcsname{112.5}
\expandafter\def\csname ppt@rev.desc.d9.uiv.natural.center-d1.opus.mean-mpix\endcsname{$+$112.5}
\expandafter\def\csname val@rev.desc.d9.uiv.natural.center-d1.opus.mean-pixels\endcsname{1125437.188}
\expandafter\def\csname cnt@rev.desc.d9.uiv.natural.center-d1.opus.mean-pixels\endcsname{1{,}125{,}437}
\expandafter\def\csname pct@rev.desc.d9.uiv.natural.center-d1.opus.mean-pixels\endcsname{112543718.8}
\expandafter\def\csname ppt@rev.desc.d9.uiv.natural.center-d1.opus.mean-pixels\endcsname{$+$112543718.8}
\expandafter\def\csname val@rev.desc.d9.uiv.natural.center-d1.opus.mean-reader-mpix\endcsname{0.539}
\expandafter\def\csname cnt@rev.desc.d9.uiv.natural.center-d1.opus.mean-reader-mpix\endcsname{1}
\expandafter\def\csname pct@rev.desc.d9.uiv.natural.center-d1.opus.mean-reader-mpix\endcsname{53.9}
\expandafter\def\csname ppt@rev.desc.d9.uiv.natural.center-d1.opus.mean-reader-mpix\endcsname{$+$53.9}
\expandafter\def\csname val@rev.desc.d9.uiv.natural.center-d1.opus.mean-requests\endcsname{2.000}
\expandafter\def\csname cnt@rev.desc.d9.uiv.natural.center-d1.opus.mean-requests\endcsname{2}
\expandafter\def\csname pct@rev.desc.d9.uiv.natural.center-d1.opus.mean-requests\endcsname{200.0}
\expandafter\def\csname ppt@rev.desc.d9.uiv.natural.center-d1.opus.mean-requests\endcsname{$+$200.0}
\expandafter\def\csname val@rev.desc.d9.uiv.natural.center-d1.opus.n-jobs\endcsname{304}
\expandafter\def\csname cnt@rev.desc.d9.uiv.natural.center-d1.opus.n-jobs\endcsname{304}
\expandafter\def\csname pct@rev.desc.d9.uiv.natural.center-d1.opus.n-jobs\endcsname{30400.0}
\expandafter\def\csname ppt@rev.desc.d9.uiv.natural.center-d1.opus.n-jobs\endcsname{$+$30400.0}
\expandafter\def\csname val@rev.desc.d9.uiv.natural.center-d1.opus.n-requests-match-scores\endcsname{1}
\expandafter\def\csname cnt@rev.desc.d9.uiv.natural.center-d1.opus.n-requests-match-scores\endcsname{1}
\expandafter\def\csname pct@rev.desc.d9.uiv.natural.center-d1.opus.n-requests-match-scores\endcsname{100.0}
\expandafter\def\csname ppt@rev.desc.d9.uiv.natural.center-d1.opus.n-requests-match-scores\endcsname{$+$100.0}
\expandafter\def\csname val@rev.desc.d9.uiv.natural.center-d1.opus.ratio-to-direct\endcsname{1.918}
\expandafter\def\csname cnt@rev.desc.d9.uiv.natural.center-d1.opus.ratio-to-direct\endcsname{2}
\expandafter\def\csname pct@rev.desc.d9.uiv.natural.center-d1.opus.ratio-to-direct\endcsname{191.8}
\expandafter\def\csname ppt@rev.desc.d9.uiv.natural.center-d1.opus.ratio-to-direct\endcsname{$+$191.8}
\expandafter\def\csname val@rev.desc.d9.uiv.natural.center-d1.ratio-to-direct\endcsname{1.916}
\expandafter\def\csname cnt@rev.desc.d9.uiv.natural.center-d1.ratio-to-direct\endcsname{2}
\expandafter\def\csname pct@rev.desc.d9.uiv.natural.center-d1.ratio-to-direct\endcsname{191.6}
\expandafter\def\csname ppt@rev.desc.d9.uiv.natural.center-d1.ratio-to-direct\endcsname{$+$191.6}
\expandafter\def\csname val@rev.desc.d9.uiv.natural.center-d1.sonnet.mean-mpix\endcsname{1.127}
\expandafter\def\csname cnt@rev.desc.d9.uiv.natural.center-d1.sonnet.mean-mpix\endcsname{1}
\expandafter\def\csname pct@rev.desc.d9.uiv.natural.center-d1.sonnet.mean-mpix\endcsname{112.7}
\expandafter\def\csname ppt@rev.desc.d9.uiv.natural.center-d1.sonnet.mean-mpix\endcsname{$+$112.7}
\expandafter\def\csname val@rev.desc.d9.uiv.natural.center-d1.sonnet.mean-pixels\endcsname{1127472.595}
\expandafter\def\csname cnt@rev.desc.d9.uiv.natural.center-d1.sonnet.mean-pixels\endcsname{1{,}127{,}473}
\expandafter\def\csname pct@rev.desc.d9.uiv.natural.center-d1.sonnet.mean-pixels\endcsname{112747259.5}
\expandafter\def\csname ppt@rev.desc.d9.uiv.natural.center-d1.sonnet.mean-pixels\endcsname{$+$112747259.5}
\expandafter\def\csname val@rev.desc.d9.uiv.natural.center-d1.sonnet.mean-reader-mpix\endcsname{0.539}
\expandafter\def\csname cnt@rev.desc.d9.uiv.natural.center-d1.sonnet.mean-reader-mpix\endcsname{1}
\expandafter\def\csname pct@rev.desc.d9.uiv.natural.center-d1.sonnet.mean-reader-mpix\endcsname{53.9}
\expandafter\def\csname ppt@rev.desc.d9.uiv.natural.center-d1.sonnet.mean-reader-mpix\endcsname{$+$53.9}
\expandafter\def\csname val@rev.desc.d9.uiv.natural.center-d1.sonnet.mean-requests\endcsname{2.003}
\expandafter\def\csname cnt@rev.desc.d9.uiv.natural.center-d1.sonnet.mean-requests\endcsname{2}
\expandafter\def\csname pct@rev.desc.d9.uiv.natural.center-d1.sonnet.mean-requests\endcsname{200.3}
\expandafter\def\csname ppt@rev.desc.d9.uiv.natural.center-d1.sonnet.mean-requests\endcsname{$+$200.3}
\expandafter\def\csname val@rev.desc.d9.uiv.natural.center-d1.sonnet.n-jobs\endcsname{304}
\expandafter\def\csname cnt@rev.desc.d9.uiv.natural.center-d1.sonnet.n-jobs\endcsname{304}
\expandafter\def\csname pct@rev.desc.d9.uiv.natural.center-d1.sonnet.n-jobs\endcsname{30400.0}
\expandafter\def\csname ppt@rev.desc.d9.uiv.natural.center-d1.sonnet.n-jobs\endcsname{$+$30400.0}
\expandafter\def\csname val@rev.desc.d9.uiv.natural.center-d1.sonnet.n-requests-match-scores\endcsname{1}
\expandafter\def\csname cnt@rev.desc.d9.uiv.natural.center-d1.sonnet.n-requests-match-scores\endcsname{1}
\expandafter\def\csname pct@rev.desc.d9.uiv.natural.center-d1.sonnet.n-requests-match-scores\endcsname{100.0}
\expandafter\def\csname ppt@rev.desc.d9.uiv.natural.center-d1.sonnet.n-requests-match-scores\endcsname{$+$100.0}
\expandafter\def\csname val@rev.desc.d9.uiv.natural.center-d1.sonnet.ratio-to-direct\endcsname{1.921}
\expandafter\def\csname cnt@rev.desc.d9.uiv.natural.center-d1.sonnet.ratio-to-direct\endcsname{2}
\expandafter\def\csname pct@rev.desc.d9.uiv.natural.center-d1.sonnet.ratio-to-direct\endcsname{192.1}
\expandafter\def\csname ppt@rev.desc.d9.uiv.natural.center-d1.sonnet.ratio-to-direct\endcsname{$+$192.1}
\expandafter\def\csname val@rev.desc.d9.uiv.natural.context-d1.haiku.mean-mpix\endcsname{1.345}
\expandafter\def\csname cnt@rev.desc.d9.uiv.natural.context-d1.haiku.mean-mpix\endcsname{1}
\expandafter\def\csname pct@rev.desc.d9.uiv.natural.context-d1.haiku.mean-mpix\endcsname{134.5}
\expandafter\def\csname ppt@rev.desc.d9.uiv.natural.context-d1.haiku.mean-mpix\endcsname{$+$134.5}
\expandafter\def\csname val@rev.desc.d9.uiv.natural.context-d1.haiku.mean-pixels\endcsname{1345003.831}
\expandafter\def\csname cnt@rev.desc.d9.uiv.natural.context-d1.haiku.mean-pixels\endcsname{1{,}345{,}004}
\expandafter\def\csname pct@rev.desc.d9.uiv.natural.context-d1.haiku.mean-pixels\endcsname{134500383.1}
\expandafter\def\csname ppt@rev.desc.d9.uiv.natural.context-d1.haiku.mean-pixels\endcsname{$+$134500383.1}
\expandafter\def\csname val@rev.desc.d9.uiv.natural.context-d1.haiku.mean-reader-mpix\endcsname{0.762}
\expandafter\def\csname cnt@rev.desc.d9.uiv.natural.context-d1.haiku.mean-reader-mpix\endcsname{1}
\expandafter\def\csname pct@rev.desc.d9.uiv.natural.context-d1.haiku.mean-reader-mpix\endcsname{76.2}
\expandafter\def\csname ppt@rev.desc.d9.uiv.natural.context-d1.haiku.mean-reader-mpix\endcsname{$+$76.2}
\expandafter\def\csname val@rev.desc.d9.uiv.natural.context-d1.haiku.mean-requests\endcsname{2.000}
\expandafter\def\csname cnt@rev.desc.d9.uiv.natural.context-d1.haiku.mean-requests\endcsname{2}
\expandafter\def\csname pct@rev.desc.d9.uiv.natural.context-d1.haiku.mean-requests\endcsname{200.0}
\expandafter\def\csname ppt@rev.desc.d9.uiv.natural.context-d1.haiku.mean-requests\endcsname{$+$200.0}
\expandafter\def\csname val@rev.desc.d9.uiv.natural.context-d1.haiku.n-jobs\endcsname{160}
\expandafter\def\csname cnt@rev.desc.d9.uiv.natural.context-d1.haiku.n-jobs\endcsname{160}
\expandafter\def\csname pct@rev.desc.d9.uiv.natural.context-d1.haiku.n-jobs\endcsname{16000.0}
\expandafter\def\csname ppt@rev.desc.d9.uiv.natural.context-d1.haiku.n-jobs\endcsname{$+$16000.0}
\expandafter\def\csname val@rev.desc.d9.uiv.natural.context-d1.haiku.n-requests-match-scores\endcsname{1}
\expandafter\def\csname cnt@rev.desc.d9.uiv.natural.context-d1.haiku.n-requests-match-scores\endcsname{1}
\expandafter\def\csname pct@rev.desc.d9.uiv.natural.context-d1.haiku.n-requests-match-scores\endcsname{100.0}
\expandafter\def\csname ppt@rev.desc.d9.uiv.natural.context-d1.haiku.n-requests-match-scores\endcsname{$+$100.0}
\expandafter\def\csname val@rev.desc.d9.uiv.natural.context-d1.haiku.ratio-to-direct\endcsname{2.292}
\expandafter\def\csname cnt@rev.desc.d9.uiv.natural.context-d1.haiku.ratio-to-direct\endcsname{2}
\expandafter\def\csname pct@rev.desc.d9.uiv.natural.context-d1.haiku.ratio-to-direct\endcsname{229.2}
\expandafter\def\csname ppt@rev.desc.d9.uiv.natural.context-d1.haiku.ratio-to-direct\endcsname{$+$229.2}
\expandafter\def\csname val@rev.desc.d9.uiv.natural.context-d1.mean-mpix\endcsname{1.353}
\expandafter\def\csname cnt@rev.desc.d9.uiv.natural.context-d1.mean-mpix\endcsname{1}
\expandafter\def\csname pct@rev.desc.d9.uiv.natural.context-d1.mean-mpix\endcsname{135.3}
\expandafter\def\csname ppt@rev.desc.d9.uiv.natural.context-d1.mean-mpix\endcsname{$+$135.3}
\expandafter\def\csname val@rev.desc.d9.uiv.natural.context-d1.mean-pixels\endcsname{1353224.958}
\expandafter\def\csname cnt@rev.desc.d9.uiv.natural.context-d1.mean-pixels\endcsname{1{,}353{,}225}
\expandafter\def\csname pct@rev.desc.d9.uiv.natural.context-d1.mean-pixels\endcsname{135322495.8}
\expandafter\def\csname ppt@rev.desc.d9.uiv.natural.context-d1.mean-pixels\endcsname{$+$135322495.8}
\expandafter\def\csname val@rev.desc.d9.uiv.natural.context-d1.mean-reader-mpix\endcsname{0.767}
\expandafter\def\csname cnt@rev.desc.d9.uiv.natural.context-d1.mean-reader-mpix\endcsname{1}
\expandafter\def\csname pct@rev.desc.d9.uiv.natural.context-d1.mean-reader-mpix\endcsname{76.7}
\expandafter\def\csname ppt@rev.desc.d9.uiv.natural.context-d1.mean-reader-mpix\endcsname{$+$76.7}
\expandafter\def\csname val@rev.desc.d9.uiv.natural.context-d1.mean-requests\endcsname{2.001}
\expandafter\def\csname cnt@rev.desc.d9.uiv.natural.context-d1.mean-requests\endcsname{2}
\expandafter\def\csname pct@rev.desc.d9.uiv.natural.context-d1.mean-requests\endcsname{200.1}
\expandafter\def\csname ppt@rev.desc.d9.uiv.natural.context-d1.mean-requests\endcsname{$+$200.1}
\expandafter\def\csname val@rev.desc.d9.uiv.natural.context-d1.n-jobs\endcsname{768}
\expandafter\def\csname cnt@rev.desc.d9.uiv.natural.context-d1.n-jobs\endcsname{768}
\expandafter\def\csname pct@rev.desc.d9.uiv.natural.context-d1.n-jobs\endcsname{76800.0}
\expandafter\def\csname ppt@rev.desc.d9.uiv.natural.context-d1.n-jobs\endcsname{$+$76800.0}
\expandafter\def\csname val@rev.desc.d9.uiv.natural.context-d1.n-requests-match-scores\endcsname{1}
\expandafter\def\csname cnt@rev.desc.d9.uiv.natural.context-d1.n-requests-match-scores\endcsname{1}
\expandafter\def\csname pct@rev.desc.d9.uiv.natural.context-d1.n-requests-match-scores\endcsname{100.0}
\expandafter\def\csname ppt@rev.desc.d9.uiv.natural.context-d1.n-requests-match-scores\endcsname{$+$100.0}
\expandafter\def\csname val@rev.desc.d9.uiv.natural.context-d1.opus.mean-mpix\endcsname{1.354}
\expandafter\def\csname cnt@rev.desc.d9.uiv.natural.context-d1.opus.mean-mpix\endcsname{1}
\expandafter\def\csname pct@rev.desc.d9.uiv.natural.context-d1.opus.mean-mpix\endcsname{135.4}
\expandafter\def\csname ppt@rev.desc.d9.uiv.natural.context-d1.opus.mean-mpix\endcsname{$+$135.4}
\expandafter\def\csname val@rev.desc.d9.uiv.natural.context-d1.opus.mean-pixels\endcsname{1354363.717}
\expandafter\def\csname cnt@rev.desc.d9.uiv.natural.context-d1.opus.mean-pixels\endcsname{1{,}354{,}364}
\expandafter\def\csname pct@rev.desc.d9.uiv.natural.context-d1.opus.mean-pixels\endcsname{135436371.7}
\expandafter\def\csname ppt@rev.desc.d9.uiv.natural.context-d1.opus.mean-pixels\endcsname{$+$135436371.7}
\expandafter\def\csname val@rev.desc.d9.uiv.natural.context-d1.opus.mean-reader-mpix\endcsname{0.768}
\expandafter\def\csname cnt@rev.desc.d9.uiv.natural.context-d1.opus.mean-reader-mpix\endcsname{1}
\expandafter\def\csname pct@rev.desc.d9.uiv.natural.context-d1.opus.mean-reader-mpix\endcsname{76.8}
\expandafter\def\csname ppt@rev.desc.d9.uiv.natural.context-d1.opus.mean-reader-mpix\endcsname{$+$76.8}
\expandafter\def\csname val@rev.desc.d9.uiv.natural.context-d1.opus.mean-requests\endcsname{2.000}
\expandafter\def\csname cnt@rev.desc.d9.uiv.natural.context-d1.opus.mean-requests\endcsname{2}
\expandafter\def\csname pct@rev.desc.d9.uiv.natural.context-d1.opus.mean-requests\endcsname{200.0}
\expandafter\def\csname ppt@rev.desc.d9.uiv.natural.context-d1.opus.mean-requests\endcsname{$+$200.0}
\expandafter\def\csname val@rev.desc.d9.uiv.natural.context-d1.opus.n-jobs\endcsname{304}
\expandafter\def\csname cnt@rev.desc.d9.uiv.natural.context-d1.opus.n-jobs\endcsname{304}
\expandafter\def\csname pct@rev.desc.d9.uiv.natural.context-d1.opus.n-jobs\endcsname{30400.0}
\expandafter\def\csname ppt@rev.desc.d9.uiv.natural.context-d1.opus.n-jobs\endcsname{$+$30400.0}
\expandafter\def\csname val@rev.desc.d9.uiv.natural.context-d1.opus.n-requests-match-scores\endcsname{1}
\expandafter\def\csname cnt@rev.desc.d9.uiv.natural.context-d1.opus.n-requests-match-scores\endcsname{1}
\expandafter\def\csname pct@rev.desc.d9.uiv.natural.context-d1.opus.n-requests-match-scores\endcsname{100.0}
\expandafter\def\csname ppt@rev.desc.d9.uiv.natural.context-d1.opus.n-requests-match-scores\endcsname{$+$100.0}
\expandafter\def\csname val@rev.desc.d9.uiv.natural.context-d1.opus.ratio-to-direct\endcsname{2.308}
\expandafter\def\csname cnt@rev.desc.d9.uiv.natural.context-d1.opus.ratio-to-direct\endcsname{2}
\expandafter\def\csname pct@rev.desc.d9.uiv.natural.context-d1.opus.ratio-to-direct\endcsname{230.8}
\expandafter\def\csname ppt@rev.desc.d9.uiv.natural.context-d1.opus.ratio-to-direct\endcsname{$+$230.8}
\expandafter\def\csname val@rev.desc.d9.uiv.natural.context-d1.ratio-to-direct\endcsname{2.306}
\expandafter\def\csname cnt@rev.desc.d9.uiv.natural.context-d1.ratio-to-direct\endcsname{2}
\expandafter\def\csname pct@rev.desc.d9.uiv.natural.context-d1.ratio-to-direct\endcsname{230.6}
\expandafter\def\csname ppt@rev.desc.d9.uiv.natural.context-d1.ratio-to-direct\endcsname{$+$230.6}
\expandafter\def\csname val@rev.desc.d9.uiv.natural.context-d1.sonnet.mean-mpix\endcsname{1.356}
\expandafter\def\csname cnt@rev.desc.d9.uiv.natural.context-d1.sonnet.mean-mpix\endcsname{1}
\expandafter\def\csname pct@rev.desc.d9.uiv.natural.context-d1.sonnet.mean-mpix\endcsname{135.6}
\expandafter\def\csname ppt@rev.desc.d9.uiv.natural.context-d1.sonnet.mean-mpix\endcsname{$+$135.6}
\expandafter\def\csname val@rev.desc.d9.uiv.natural.context-d1.sonnet.mean-pixels\endcsname{1356413.109}
\expandafter\def\csname cnt@rev.desc.d9.uiv.natural.context-d1.sonnet.mean-pixels\endcsname{1{,}356{,}413}
\expandafter\def\csname pct@rev.desc.d9.uiv.natural.context-d1.sonnet.mean-pixels\endcsname{135641310.9}
\expandafter\def\csname ppt@rev.desc.d9.uiv.natural.context-d1.sonnet.mean-pixels\endcsname{$+$135641310.9}
\expandafter\def\csname val@rev.desc.d9.uiv.natural.context-d1.sonnet.mean-reader-mpix\endcsname{0.768}
\expandafter\def\csname cnt@rev.desc.d9.uiv.natural.context-d1.sonnet.mean-reader-mpix\endcsname{1}
\expandafter\def\csname pct@rev.desc.d9.uiv.natural.context-d1.sonnet.mean-reader-mpix\endcsname{76.8}
\expandafter\def\csname ppt@rev.desc.d9.uiv.natural.context-d1.sonnet.mean-reader-mpix\endcsname{$+$76.8}
\expandafter\def\csname val@rev.desc.d9.uiv.natural.context-d1.sonnet.mean-requests\endcsname{2.003}
\expandafter\def\csname cnt@rev.desc.d9.uiv.natural.context-d1.sonnet.mean-requests\endcsname{2}
\expandafter\def\csname pct@rev.desc.d9.uiv.natural.context-d1.sonnet.mean-requests\endcsname{200.3}
\expandafter\def\csname ppt@rev.desc.d9.uiv.natural.context-d1.sonnet.mean-requests\endcsname{$+$200.3}
\expandafter\def\csname val@rev.desc.d9.uiv.natural.context-d1.sonnet.n-jobs\endcsname{304}
\expandafter\def\csname cnt@rev.desc.d9.uiv.natural.context-d1.sonnet.n-jobs\endcsname{304}
\expandafter\def\csname pct@rev.desc.d9.uiv.natural.context-d1.sonnet.n-jobs\endcsname{30400.0}
\expandafter\def\csname ppt@rev.desc.d9.uiv.natural.context-d1.sonnet.n-jobs\endcsname{$+$30400.0}
\expandafter\def\csname val@rev.desc.d9.uiv.natural.context-d1.sonnet.n-requests-match-scores\endcsname{1}
\expandafter\def\csname cnt@rev.desc.d9.uiv.natural.context-d1.sonnet.n-requests-match-scores\endcsname{1}
\expandafter\def\csname pct@rev.desc.d9.uiv.natural.context-d1.sonnet.n-requests-match-scores\endcsname{100.0}
\expandafter\def\csname ppt@rev.desc.d9.uiv.natural.context-d1.sonnet.n-requests-match-scores\endcsname{$+$100.0}
\expandafter\def\csname val@rev.desc.d9.uiv.natural.context-d1.sonnet.ratio-to-direct\endcsname{2.312}
\expandafter\def\csname cnt@rev.desc.d9.uiv.natural.context-d1.sonnet.ratio-to-direct\endcsname{2}
\expandafter\def\csname pct@rev.desc.d9.uiv.natural.context-d1.sonnet.ratio-to-direct\endcsname{231.2}
\expandafter\def\csname ppt@rev.desc.d9.uiv.natural.context-d1.sonnet.ratio-to-direct\endcsname{$+$231.2}
\expandafter\def\csname val@rev.desc.d9.uiv.natural.dual-d2.haiku.mean-mpix\endcsname{0.951}
\expandafter\def\csname cnt@rev.desc.d9.uiv.natural.dual-d2.haiku.mean-mpix\endcsname{1}
\expandafter\def\csname pct@rev.desc.d9.uiv.natural.dual-d2.haiku.mean-mpix\endcsname{95.1}
\expandafter\def\csname ppt@rev.desc.d9.uiv.natural.dual-d2.haiku.mean-mpix\endcsname{$+$95.1}
\expandafter\def\csname val@rev.desc.d9.uiv.natural.dual-d2.haiku.mean-pixels\endcsname{951353.912}
\expandafter\def\csname cnt@rev.desc.d9.uiv.natural.dual-d2.haiku.mean-pixels\endcsname{951{,}354}
\expandafter\def\csname pct@rev.desc.d9.uiv.natural.dual-d2.haiku.mean-pixels\endcsname{95135391.2}
\expandafter\def\csname ppt@rev.desc.d9.uiv.natural.dual-d2.haiku.mean-pixels\endcsname{$+$95135391.2}
\expandafter\def\csname val@rev.desc.d9.uiv.natural.dual-d2.haiku.mean-reader-mpix\endcsname{0.283}
\expandafter\def\csname cnt@rev.desc.d9.uiv.natural.dual-d2.haiku.mean-reader-mpix\endcsname{0}
\expandafter\def\csname pct@rev.desc.d9.uiv.natural.dual-d2.haiku.mean-reader-mpix\endcsname{28.3}
\expandafter\def\csname ppt@rev.desc.d9.uiv.natural.dual-d2.haiku.mean-reader-mpix\endcsname{$+$28.3}
\expandafter\def\csname val@rev.desc.d9.uiv.natural.dual-d2.haiku.mean-requests\endcsname{3.013}
\expandafter\def\csname cnt@rev.desc.d9.uiv.natural.dual-d2.haiku.mean-requests\endcsname{3}
\expandafter\def\csname pct@rev.desc.d9.uiv.natural.dual-d2.haiku.mean-requests\endcsname{301.2}
\expandafter\def\csname ppt@rev.desc.d9.uiv.natural.dual-d2.haiku.mean-requests\endcsname{$+$301.2}
\expandafter\def\csname val@rev.desc.d9.uiv.natural.dual-d2.haiku.n-jobs\endcsname{160}
\expandafter\def\csname cnt@rev.desc.d9.uiv.natural.dual-d2.haiku.n-jobs\endcsname{160}
\expandafter\def\csname pct@rev.desc.d9.uiv.natural.dual-d2.haiku.n-jobs\endcsname{16000.0}
\expandafter\def\csname ppt@rev.desc.d9.uiv.natural.dual-d2.haiku.n-jobs\endcsname{$+$16000.0}
\expandafter\def\csname val@rev.desc.d9.uiv.natural.dual-d2.haiku.n-requests-match-scores\endcsname{1}
\expandafter\def\csname cnt@rev.desc.d9.uiv.natural.dual-d2.haiku.n-requests-match-scores\endcsname{1}
\expandafter\def\csname pct@rev.desc.d9.uiv.natural.dual-d2.haiku.n-requests-match-scores\endcsname{100.0}
\expandafter\def\csname ppt@rev.desc.d9.uiv.natural.dual-d2.haiku.n-requests-match-scores\endcsname{$+$100.0}
\expandafter\def\csname val@rev.desc.d9.uiv.natural.dual-d2.haiku.ratio-to-direct\endcsname{1.621}
\expandafter\def\csname cnt@rev.desc.d9.uiv.natural.dual-d2.haiku.ratio-to-direct\endcsname{2}
\expandafter\def\csname pct@rev.desc.d9.uiv.natural.dual-d2.haiku.ratio-to-direct\endcsname{162.1}
\expandafter\def\csname ppt@rev.desc.d9.uiv.natural.dual-d2.haiku.ratio-to-direct\endcsname{$+$162.1}
\expandafter\def\csname val@rev.desc.d9.uiv.natural.dual-d2.mean-mpix\endcsname{0.960}
\expandafter\def\csname cnt@rev.desc.d9.uiv.natural.dual-d2.mean-mpix\endcsname{1}
\expandafter\def\csname pct@rev.desc.d9.uiv.natural.dual-d2.mean-mpix\endcsname{96.0}
\expandafter\def\csname ppt@rev.desc.d9.uiv.natural.dual-d2.mean-mpix\endcsname{$+$96.0}
\expandafter\def\csname val@rev.desc.d9.uiv.natural.dual-d2.mean-pixels\endcsname{959980.257}
\expandafter\def\csname cnt@rev.desc.d9.uiv.natural.dual-d2.mean-pixels\endcsname{959{,}980}
\expandafter\def\csname pct@rev.desc.d9.uiv.natural.dual-d2.mean-pixels\endcsname{95998025.7}
\expandafter\def\csname ppt@rev.desc.d9.uiv.natural.dual-d2.mean-pixels\endcsname{$+$95998025.7}
\expandafter\def\csname val@rev.desc.d9.uiv.natural.dual-d2.mean-reader-mpix\endcsname{0.285}
\expandafter\def\csname cnt@rev.desc.d9.uiv.natural.dual-d2.mean-reader-mpix\endcsname{0}
\expandafter\def\csname pct@rev.desc.d9.uiv.natural.dual-d2.mean-reader-mpix\endcsname{28.5}
\expandafter\def\csname ppt@rev.desc.d9.uiv.natural.dual-d2.mean-reader-mpix\endcsname{$+$28.5}
\expandafter\def\csname val@rev.desc.d9.uiv.natural.dual-d2.mean-requests\endcsname{3.013}
\expandafter\def\csname cnt@rev.desc.d9.uiv.natural.dual-d2.mean-requests\endcsname{3}
\expandafter\def\csname pct@rev.desc.d9.uiv.natural.dual-d2.mean-requests\endcsname{301.3}
\expandafter\def\csname ppt@rev.desc.d9.uiv.natural.dual-d2.mean-requests\endcsname{$+$301.3}
\expandafter\def\csname val@rev.desc.d9.uiv.natural.dual-d2.n-jobs\endcsname{768}
\expandafter\def\csname cnt@rev.desc.d9.uiv.natural.dual-d2.n-jobs\endcsname{768}
\expandafter\def\csname pct@rev.desc.d9.uiv.natural.dual-d2.n-jobs\endcsname{76800.0}
\expandafter\def\csname ppt@rev.desc.d9.uiv.natural.dual-d2.n-jobs\endcsname{$+$76800.0}
\expandafter\def\csname val@rev.desc.d9.uiv.natural.dual-d2.n-requests-match-scores\endcsname{1}
\expandafter\def\csname cnt@rev.desc.d9.uiv.natural.dual-d2.n-requests-match-scores\endcsname{1}
\expandafter\def\csname pct@rev.desc.d9.uiv.natural.dual-d2.n-requests-match-scores\endcsname{100.0}
\expandafter\def\csname ppt@rev.desc.d9.uiv.natural.dual-d2.n-requests-match-scores\endcsname{$+$100.0}
\expandafter\def\csname val@rev.desc.d9.uiv.natural.dual-d2.opus.mean-mpix\endcsname{0.928}
\expandafter\def\csname cnt@rev.desc.d9.uiv.natural.dual-d2.opus.mean-mpix\endcsname{1}
\expandafter\def\csname pct@rev.desc.d9.uiv.natural.dual-d2.opus.mean-mpix\endcsname{92.8}
\expandafter\def\csname ppt@rev.desc.d9.uiv.natural.dual-d2.opus.mean-mpix\endcsname{$+$92.8}
\expandafter\def\csname val@rev.desc.d9.uiv.natural.dual-d2.opus.mean-pixels\endcsname{927638.451}
\expandafter\def\csname cnt@rev.desc.d9.uiv.natural.dual-d2.opus.mean-pixels\endcsname{927{,}638}
\expandafter\def\csname pct@rev.desc.d9.uiv.natural.dual-d2.opus.mean-pixels\endcsname{92763845.1}
\expandafter\def\csname ppt@rev.desc.d9.uiv.natural.dual-d2.opus.mean-pixels\endcsname{$+$92763845.1}
\expandafter\def\csname val@rev.desc.d9.uiv.natural.dual-d2.opus.mean-reader-mpix\endcsname{0.279}
\expandafter\def\csname cnt@rev.desc.d9.uiv.natural.dual-d2.opus.mean-reader-mpix\endcsname{0}
\expandafter\def\csname pct@rev.desc.d9.uiv.natural.dual-d2.opus.mean-reader-mpix\endcsname{27.9}
\expandafter\def\csname ppt@rev.desc.d9.uiv.natural.dual-d2.opus.mean-reader-mpix\endcsname{$+$27.9}
\expandafter\def\csname val@rev.desc.d9.uiv.natural.dual-d2.opus.mean-requests\endcsname{3.000}
\expandafter\def\csname cnt@rev.desc.d9.uiv.natural.dual-d2.opus.mean-requests\endcsname{3}
\expandafter\def\csname pct@rev.desc.d9.uiv.natural.dual-d2.opus.mean-requests\endcsname{300.0}
\expandafter\def\csname ppt@rev.desc.d9.uiv.natural.dual-d2.opus.mean-requests\endcsname{$+$300.0}
\expandafter\def\csname val@rev.desc.d9.uiv.natural.dual-d2.opus.n-jobs\endcsname{304}
\expandafter\def\csname cnt@rev.desc.d9.uiv.natural.dual-d2.opus.n-jobs\endcsname{304}
\expandafter\def\csname pct@rev.desc.d9.uiv.natural.dual-d2.opus.n-jobs\endcsname{30400.0}
\expandafter\def\csname ppt@rev.desc.d9.uiv.natural.dual-d2.opus.n-jobs\endcsname{$+$30400.0}
\expandafter\def\csname val@rev.desc.d9.uiv.natural.dual-d2.opus.n-requests-match-scores\endcsname{1}
\expandafter\def\csname cnt@rev.desc.d9.uiv.natural.dual-d2.opus.n-requests-match-scores\endcsname{1}
\expandafter\def\csname pct@rev.desc.d9.uiv.natural.dual-d2.opus.n-requests-match-scores\endcsname{100.0}
\expandafter\def\csname ppt@rev.desc.d9.uiv.natural.dual-d2.opus.n-requests-match-scores\endcsname{$+$100.0}
\expandafter\def\csname val@rev.desc.d9.uiv.natural.dual-d2.opus.ratio-to-direct\endcsname{1.581}
\expandafter\def\csname cnt@rev.desc.d9.uiv.natural.dual-d2.opus.ratio-to-direct\endcsname{2}
\expandafter\def\csname pct@rev.desc.d9.uiv.natural.dual-d2.opus.ratio-to-direct\endcsname{158.1}
\expandafter\def\csname ppt@rev.desc.d9.uiv.natural.dual-d2.opus.ratio-to-direct\endcsname{$+$158.1}
\expandafter\def\csname val@rev.desc.d9.uiv.natural.dual-d2.ratio-to-direct\endcsname{1.636}
\expandafter\def\csname cnt@rev.desc.d9.uiv.natural.dual-d2.ratio-to-direct\endcsname{2}
\expandafter\def\csname pct@rev.desc.d9.uiv.natural.dual-d2.ratio-to-direct\endcsname{163.6}
\expandafter\def\csname ppt@rev.desc.d9.uiv.natural.dual-d2.ratio-to-direct\endcsname{$+$163.6}
\expandafter\def\csname val@rev.desc.d9.uiv.natural.dual-d2.sonnet.mean-mpix\endcsname{0.997}
\expandafter\def\csname cnt@rev.desc.d9.uiv.natural.dual-d2.sonnet.mean-mpix\endcsname{1}
\expandafter\def\csname pct@rev.desc.d9.uiv.natural.dual-d2.sonnet.mean-mpix\endcsname{99.7}
\expandafter\def\csname ppt@rev.desc.d9.uiv.natural.dual-d2.sonnet.mean-mpix\endcsname{$+$99.7}
\expandafter\def\csname val@rev.desc.d9.uiv.natural.dual-d2.sonnet.mean-pixels\endcsname{996862.243}
\expandafter\def\csname cnt@rev.desc.d9.uiv.natural.dual-d2.sonnet.mean-pixels\endcsname{996{,}862}
\expandafter\def\csname pct@rev.desc.d9.uiv.natural.dual-d2.sonnet.mean-pixels\endcsname{99686224.3}
\expandafter\def\csname ppt@rev.desc.d9.uiv.natural.dual-d2.sonnet.mean-pixels\endcsname{$+$99686224.3}
\expandafter\def\csname val@rev.desc.d9.uiv.natural.dual-d2.sonnet.mean-reader-mpix\endcsname{0.293}
\expandafter\def\csname cnt@rev.desc.d9.uiv.natural.dual-d2.sonnet.mean-reader-mpix\endcsname{0}
\expandafter\def\csname pct@rev.desc.d9.uiv.natural.dual-d2.sonnet.mean-reader-mpix\endcsname{29.3}
\expandafter\def\csname ppt@rev.desc.d9.uiv.natural.dual-d2.sonnet.mean-reader-mpix\endcsname{$+$29.3}
\expandafter\def\csname val@rev.desc.d9.uiv.natural.dual-d2.sonnet.mean-requests\endcsname{3.026}
\expandafter\def\csname cnt@rev.desc.d9.uiv.natural.dual-d2.sonnet.mean-requests\endcsname{3}
\expandafter\def\csname pct@rev.desc.d9.uiv.natural.dual-d2.sonnet.mean-requests\endcsname{302.6}
\expandafter\def\csname ppt@rev.desc.d9.uiv.natural.dual-d2.sonnet.mean-requests\endcsname{$+$302.6}
\expandafter\def\csname val@rev.desc.d9.uiv.natural.dual-d2.sonnet.n-jobs\endcsname{304}
\expandafter\def\csname cnt@rev.desc.d9.uiv.natural.dual-d2.sonnet.n-jobs\endcsname{304}
\expandafter\def\csname pct@rev.desc.d9.uiv.natural.dual-d2.sonnet.n-jobs\endcsname{30400.0}
\expandafter\def\csname ppt@rev.desc.d9.uiv.natural.dual-d2.sonnet.n-jobs\endcsname{$+$30400.0}
\expandafter\def\csname val@rev.desc.d9.uiv.natural.dual-d2.sonnet.n-requests-match-scores\endcsname{1}
\expandafter\def\csname cnt@rev.desc.d9.uiv.natural.dual-d2.sonnet.n-requests-match-scores\endcsname{1}
\expandafter\def\csname pct@rev.desc.d9.uiv.natural.dual-d2.sonnet.n-requests-match-scores\endcsname{100.0}
\expandafter\def\csname ppt@rev.desc.d9.uiv.natural.dual-d2.sonnet.n-requests-match-scores\endcsname{$+$100.0}
\expandafter\def\csname val@rev.desc.d9.uiv.natural.dual-d2.sonnet.ratio-to-direct\endcsname{1.699}
\expandafter\def\csname cnt@rev.desc.d9.uiv.natural.dual-d2.sonnet.ratio-to-direct\endcsname{2}
\expandafter\def\csname pct@rev.desc.d9.uiv.natural.dual-d2.sonnet.ratio-to-direct\endcsname{169.9}
\expandafter\def\csname ppt@rev.desc.d9.uiv.natural.dual-d2.sonnet.ratio-to-direct\endcsname{$+$169.9}
\expandafter\def\csname val@rev.desc.d9.uiv.natural.grid2-d3.haiku.mean-mpix\endcsname{3.333}
\expandafter\def\csname cnt@rev.desc.d9.uiv.natural.grid2-d3.haiku.mean-mpix\endcsname{3}
\expandafter\def\csname pct@rev.desc.d9.uiv.natural.grid2-d3.haiku.mean-mpix\endcsname{333.3}
\expandafter\def\csname ppt@rev.desc.d9.uiv.natural.grid2-d3.haiku.mean-mpix\endcsname{$+$333.3}
\expandafter\def\csname val@rev.desc.d9.uiv.natural.grid2-d3.haiku.mean-pixels\endcsname{3333364.144}
\expandafter\def\csname cnt@rev.desc.d9.uiv.natural.grid2-d3.haiku.mean-pixels\endcsname{3{,}333{,}364}
\expandafter\def\csname pct@rev.desc.d9.uiv.natural.grid2-d3.haiku.mean-pixels\endcsname{333336414.4}
\expandafter\def\csname ppt@rev.desc.d9.uiv.natural.grid2-d3.haiku.mean-pixels\endcsname{$+$333336414.4}
\expandafter\def\csname val@rev.desc.d9.uiv.natural.grid2-d3.haiku.mean-reader-mpix\endcsname{0.186}
\expandafter\def\csname cnt@rev.desc.d9.uiv.natural.grid2-d3.haiku.mean-reader-mpix\endcsname{0}
\expandafter\def\csname pct@rev.desc.d9.uiv.natural.grid2-d3.haiku.mean-reader-mpix\endcsname{18.6}
\expandafter\def\csname ppt@rev.desc.d9.uiv.natural.grid2-d3.haiku.mean-reader-mpix\endcsname{$+$18.6}
\expandafter\def\csname val@rev.desc.d9.uiv.natural.grid2-d3.haiku.mean-requests\endcsname{3.969}
\expandafter\def\csname cnt@rev.desc.d9.uiv.natural.grid2-d3.haiku.mean-requests\endcsname{4}
\expandafter\def\csname pct@rev.desc.d9.uiv.natural.grid2-d3.haiku.mean-requests\endcsname{396.9}
\expandafter\def\csname ppt@rev.desc.d9.uiv.natural.grid2-d3.haiku.mean-requests\endcsname{$+$396.9}
\expandafter\def\csname val@rev.desc.d9.uiv.natural.grid2-d3.haiku.n-jobs\endcsname{160}
\expandafter\def\csname cnt@rev.desc.d9.uiv.natural.grid2-d3.haiku.n-jobs\endcsname{160}
\expandafter\def\csname pct@rev.desc.d9.uiv.natural.grid2-d3.haiku.n-jobs\endcsname{16000.0}
\expandafter\def\csname ppt@rev.desc.d9.uiv.natural.grid2-d3.haiku.n-jobs\endcsname{$+$16000.0}
\expandafter\def\csname val@rev.desc.d9.uiv.natural.grid2-d3.haiku.n-requests-match-scores\endcsname{1}
\expandafter\def\csname cnt@rev.desc.d9.uiv.natural.grid2-d3.haiku.n-requests-match-scores\endcsname{1}
\expandafter\def\csname pct@rev.desc.d9.uiv.natural.grid2-d3.haiku.n-requests-match-scores\endcsname{100.0}
\expandafter\def\csname ppt@rev.desc.d9.uiv.natural.grid2-d3.haiku.n-requests-match-scores\endcsname{$+$100.0}
\expandafter\def\csname val@rev.desc.d9.uiv.natural.grid2-d3.haiku.ratio-to-direct\endcsname{5.681}
\expandafter\def\csname cnt@rev.desc.d9.uiv.natural.grid2-d3.haiku.ratio-to-direct\endcsname{6}
\expandafter\def\csname pct@rev.desc.d9.uiv.natural.grid2-d3.haiku.ratio-to-direct\endcsname{568.1}
\expandafter\def\csname ppt@rev.desc.d9.uiv.natural.grid2-d3.haiku.ratio-to-direct\endcsname{$+$568.1}
\expandafter\def\csname val@rev.desc.d9.uiv.natural.grid2-d3.mean-mpix\endcsname{2.990}
\expandafter\def\csname cnt@rev.desc.d9.uiv.natural.grid2-d3.mean-mpix\endcsname{3}
\expandafter\def\csname pct@rev.desc.d9.uiv.natural.grid2-d3.mean-mpix\endcsname{299.0}
\expandafter\def\csname ppt@rev.desc.d9.uiv.natural.grid2-d3.mean-mpix\endcsname{$+$299.0}
\expandafter\def\csname val@rev.desc.d9.uiv.natural.grid2-d3.mean-pixels\endcsname{2989906.417}
\expandafter\def\csname cnt@rev.desc.d9.uiv.natural.grid2-d3.mean-pixels\endcsname{2{,}989{,}906}
\expandafter\def\csname pct@rev.desc.d9.uiv.natural.grid2-d3.mean-pixels\endcsname{298990641.7}
\expandafter\def\csname ppt@rev.desc.d9.uiv.natural.grid2-d3.mean-pixels\endcsname{$+$298990641.7}
\expandafter\def\csname val@rev.desc.d9.uiv.natural.grid2-d3.mean-reader-mpix\endcsname{0.233}
\expandafter\def\csname cnt@rev.desc.d9.uiv.natural.grid2-d3.mean-reader-mpix\endcsname{0}
\expandafter\def\csname pct@rev.desc.d9.uiv.natural.grid2-d3.mean-reader-mpix\endcsname{23.3}
\expandafter\def\csname ppt@rev.desc.d9.uiv.natural.grid2-d3.mean-reader-mpix\endcsname{$+$23.3}
\expandafter\def\csname val@rev.desc.d9.uiv.natural.grid2-d3.mean-requests\endcsname{3.689}
\expandafter\def\csname cnt@rev.desc.d9.uiv.natural.grid2-d3.mean-requests\endcsname{4}
\expandafter\def\csname pct@rev.desc.d9.uiv.natural.grid2-d3.mean-requests\endcsname{368.9}
\expandafter\def\csname ppt@rev.desc.d9.uiv.natural.grid2-d3.mean-requests\endcsname{$+$368.9}
\expandafter\def\csname val@rev.desc.d9.uiv.natural.grid2-d3.n-jobs\endcsname{768}
\expandafter\def\csname cnt@rev.desc.d9.uiv.natural.grid2-d3.n-jobs\endcsname{768}
\expandafter\def\csname pct@rev.desc.d9.uiv.natural.grid2-d3.n-jobs\endcsname{76800.0}
\expandafter\def\csname ppt@rev.desc.d9.uiv.natural.grid2-d3.n-jobs\endcsname{$+$76800.0}
\expandafter\def\csname val@rev.desc.d9.uiv.natural.grid2-d3.n-requests-match-scores\endcsname{1}
\expandafter\def\csname cnt@rev.desc.d9.uiv.natural.grid2-d3.n-requests-match-scores\endcsname{1}
\expandafter\def\csname pct@rev.desc.d9.uiv.natural.grid2-d3.n-requests-match-scores\endcsname{100.0}
\expandafter\def\csname ppt@rev.desc.d9.uiv.natural.grid2-d3.n-requests-match-scores\endcsname{$+$100.0}
\expandafter\def\csname val@rev.desc.d9.uiv.natural.grid2-d3.opus.mean-mpix\endcsname{2.796}
\expandafter\def\csname cnt@rev.desc.d9.uiv.natural.grid2-d3.opus.mean-mpix\endcsname{3}
\expandafter\def\csname pct@rev.desc.d9.uiv.natural.grid2-d3.opus.mean-mpix\endcsname{279.6}
\expandafter\def\csname ppt@rev.desc.d9.uiv.natural.grid2-d3.opus.mean-mpix\endcsname{$+$279.6}
\expandafter\def\csname val@rev.desc.d9.uiv.natural.grid2-d3.opus.mean-pixels\endcsname{2796211.859}
\expandafter\def\csname cnt@rev.desc.d9.uiv.natural.grid2-d3.opus.mean-pixels\endcsname{2{,}796{,}212}
\expandafter\def\csname pct@rev.desc.d9.uiv.natural.grid2-d3.opus.mean-pixels\endcsname{279621185.9}
\expandafter\def\csname ppt@rev.desc.d9.uiv.natural.grid2-d3.opus.mean-pixels\endcsname{$+$279621185.9}
\expandafter\def\csname val@rev.desc.d9.uiv.natural.grid2-d3.opus.mean-reader-mpix\endcsname{0.261}
\expandafter\def\csname cnt@rev.desc.d9.uiv.natural.grid2-d3.opus.mean-reader-mpix\endcsname{0}
\expandafter\def\csname pct@rev.desc.d9.uiv.natural.grid2-d3.opus.mean-reader-mpix\endcsname{26.1}
\expandafter\def\csname ppt@rev.desc.d9.uiv.natural.grid2-d3.opus.mean-reader-mpix\endcsname{$+$26.1}
\expandafter\def\csname val@rev.desc.d9.uiv.natural.grid2-d3.opus.mean-requests\endcsname{3.533}
\expandafter\def\csname cnt@rev.desc.d9.uiv.natural.grid2-d3.opus.mean-requests\endcsname{4}
\expandafter\def\csname pct@rev.desc.d9.uiv.natural.grid2-d3.opus.mean-requests\endcsname{353.3}
\expandafter\def\csname ppt@rev.desc.d9.uiv.natural.grid2-d3.opus.mean-requests\endcsname{$+$353.3}
\expandafter\def\csname val@rev.desc.d9.uiv.natural.grid2-d3.opus.n-jobs\endcsname{304}
\expandafter\def\csname cnt@rev.desc.d9.uiv.natural.grid2-d3.opus.n-jobs\endcsname{304}
\expandafter\def\csname pct@rev.desc.d9.uiv.natural.grid2-d3.opus.n-jobs\endcsname{30400.0}
\expandafter\def\csname ppt@rev.desc.d9.uiv.natural.grid2-d3.opus.n-jobs\endcsname{$+$30400.0}
\expandafter\def\csname val@rev.desc.d9.uiv.natural.grid2-d3.opus.n-requests-match-scores\endcsname{1}
\expandafter\def\csname cnt@rev.desc.d9.uiv.natural.grid2-d3.opus.n-requests-match-scores\endcsname{1}
\expandafter\def\csname pct@rev.desc.d9.uiv.natural.grid2-d3.opus.n-requests-match-scores\endcsname{100.0}
\expandafter\def\csname ppt@rev.desc.d9.uiv.natural.grid2-d3.opus.n-requests-match-scores\endcsname{$+$100.0}
\expandafter\def\csname val@rev.desc.d9.uiv.natural.grid2-d3.opus.ratio-to-direct\endcsname{4.765}
\expandafter\def\csname cnt@rev.desc.d9.uiv.natural.grid2-d3.opus.ratio-to-direct\endcsname{5}
\expandafter\def\csname pct@rev.desc.d9.uiv.natural.grid2-d3.opus.ratio-to-direct\endcsname{476.5}
\expandafter\def\csname ppt@rev.desc.d9.uiv.natural.grid2-d3.opus.ratio-to-direct\endcsname{$+$476.5}
\expandafter\def\csname val@rev.desc.d9.uiv.natural.grid2-d3.ratio-to-direct\endcsname{5.095}
\expandafter\def\csname cnt@rev.desc.d9.uiv.natural.grid2-d3.ratio-to-direct\endcsname{5}
\expandafter\def\csname pct@rev.desc.d9.uiv.natural.grid2-d3.ratio-to-direct\endcsname{509.5}
\expandafter\def\csname ppt@rev.desc.d9.uiv.natural.grid2-d3.ratio-to-direct\endcsname{$+$509.5}
\expandafter\def\csname val@rev.desc.d9.uiv.natural.grid2-d3.sonnet.mean-mpix\endcsname{3.003}
\expandafter\def\csname cnt@rev.desc.d9.uiv.natural.grid2-d3.sonnet.mean-mpix\endcsname{3}
\expandafter\def\csname pct@rev.desc.d9.uiv.natural.grid2-d3.sonnet.mean-mpix\endcsname{300.3}
\expandafter\def\csname ppt@rev.desc.d9.uiv.natural.grid2-d3.sonnet.mean-mpix\endcsname{$+$300.3}
\expandafter\def\csname val@rev.desc.d9.uiv.natural.grid2-d3.sonnet.mean-pixels\endcsname{3002833.750}
\expandafter\def\csname cnt@rev.desc.d9.uiv.natural.grid2-d3.sonnet.mean-pixels\endcsname{3{,}002{,}834}
\expandafter\def\csname pct@rev.desc.d9.uiv.natural.grid2-d3.sonnet.mean-pixels\endcsname{300283375.0}
\expandafter\def\csname ppt@rev.desc.d9.uiv.natural.grid2-d3.sonnet.mean-pixels\endcsname{$+$300283375.0}
\expandafter\def\csname val@rev.desc.d9.uiv.natural.grid2-d3.sonnet.mean-reader-mpix\endcsname{0.229}
\expandafter\def\csname cnt@rev.desc.d9.uiv.natural.grid2-d3.sonnet.mean-reader-mpix\endcsname{0}
\expandafter\def\csname pct@rev.desc.d9.uiv.natural.grid2-d3.sonnet.mean-reader-mpix\endcsname{22.9}
\expandafter\def\csname ppt@rev.desc.d9.uiv.natural.grid2-d3.sonnet.mean-reader-mpix\endcsname{$+$22.9}
\expandafter\def\csname val@rev.desc.d9.uiv.natural.grid2-d3.sonnet.mean-requests\endcsname{3.697}
\expandafter\def\csname cnt@rev.desc.d9.uiv.natural.grid2-d3.sonnet.mean-requests\endcsname{4}
\expandafter\def\csname pct@rev.desc.d9.uiv.natural.grid2-d3.sonnet.mean-requests\endcsname{369.7}
\expandafter\def\csname ppt@rev.desc.d9.uiv.natural.grid2-d3.sonnet.mean-requests\endcsname{$+$369.7}
\expandafter\def\csname val@rev.desc.d9.uiv.natural.grid2-d3.sonnet.n-jobs\endcsname{304}
\expandafter\def\csname cnt@rev.desc.d9.uiv.natural.grid2-d3.sonnet.n-jobs\endcsname{304}
\expandafter\def\csname pct@rev.desc.d9.uiv.natural.grid2-d3.sonnet.n-jobs\endcsname{30400.0}
\expandafter\def\csname ppt@rev.desc.d9.uiv.natural.grid2-d3.sonnet.n-jobs\endcsname{$+$30400.0}
\expandafter\def\csname val@rev.desc.d9.uiv.natural.grid2-d3.sonnet.n-requests-match-scores\endcsname{1}
\expandafter\def\csname cnt@rev.desc.d9.uiv.natural.grid2-d3.sonnet.n-requests-match-scores\endcsname{1}
\expandafter\def\csname pct@rev.desc.d9.uiv.natural.grid2-d3.sonnet.n-requests-match-scores\endcsname{100.0}
\expandafter\def\csname ppt@rev.desc.d9.uiv.natural.grid2-d3.sonnet.n-requests-match-scores\endcsname{$+$100.0}
\expandafter\def\csname val@rev.desc.d9.uiv.natural.grid2-d3.sonnet.ratio-to-direct\endcsname{5.117}
\expandafter\def\csname cnt@rev.desc.d9.uiv.natural.grid2-d3.sonnet.ratio-to-direct\endcsname{5}
\expandafter\def\csname pct@rev.desc.d9.uiv.natural.grid2-d3.sonnet.ratio-to-direct\endcsname{511.7}
\expandafter\def\csname ppt@rev.desc.d9.uiv.natural.grid2-d3.sonnet.ratio-to-direct\endcsname{$+$511.7}
\expandafter\def\csname val@rev.desc.d9.uiv.natural.zoomclick-d3.haiku.mean-mpix\endcsname{1.384}
\expandafter\def\csname cnt@rev.desc.d9.uiv.natural.zoomclick-d3.haiku.mean-mpix\endcsname{1}
\expandafter\def\csname pct@rev.desc.d9.uiv.natural.zoomclick-d3.haiku.mean-mpix\endcsname{138.4}
\expandafter\def\csname ppt@rev.desc.d9.uiv.natural.zoomclick-d3.haiku.mean-mpix\endcsname{$+$138.4}
\expandafter\def\csname val@rev.desc.d9.uiv.natural.zoomclick-d3.haiku.mean-pixels\endcsname{1383573.944}
\expandafter\def\csname cnt@rev.desc.d9.uiv.natural.zoomclick-d3.haiku.mean-pixels\endcsname{1{,}383{,}574}
\expandafter\def\csname pct@rev.desc.d9.uiv.natural.zoomclick-d3.haiku.mean-pixels\endcsname{138357394.4}
\expandafter\def\csname ppt@rev.desc.d9.uiv.natural.zoomclick-d3.haiku.mean-pixels\endcsname{$+$138357394.4}
\expandafter\def\csname val@rev.desc.d9.uiv.natural.zoomclick-d3.haiku.mean-reader-mpix\endcsname{0.374}
\expandafter\def\csname cnt@rev.desc.d9.uiv.natural.zoomclick-d3.haiku.mean-reader-mpix\endcsname{0}
\expandafter\def\csname pct@rev.desc.d9.uiv.natural.zoomclick-d3.haiku.mean-reader-mpix\endcsname{37.4}
\expandafter\def\csname ppt@rev.desc.d9.uiv.natural.zoomclick-d3.haiku.mean-reader-mpix\endcsname{$+$37.4}
\expandafter\def\csname val@rev.desc.d9.uiv.natural.zoomclick-d3.haiku.mean-requests\endcsname{2.975}
\expandafter\def\csname cnt@rev.desc.d9.uiv.natural.zoomclick-d3.haiku.mean-requests\endcsname{3}
\expandafter\def\csname pct@rev.desc.d9.uiv.natural.zoomclick-d3.haiku.mean-requests\endcsname{297.5}
\expandafter\def\csname ppt@rev.desc.d9.uiv.natural.zoomclick-d3.haiku.mean-requests\endcsname{$+$297.5}
\expandafter\def\csname val@rev.desc.d9.uiv.natural.zoomclick-d3.haiku.n-jobs\endcsname{160}
\expandafter\def\csname cnt@rev.desc.d9.uiv.natural.zoomclick-d3.haiku.n-jobs\endcsname{160}
\expandafter\def\csname pct@rev.desc.d9.uiv.natural.zoomclick-d3.haiku.n-jobs\endcsname{16000.0}
\expandafter\def\csname ppt@rev.desc.d9.uiv.natural.zoomclick-d3.haiku.n-jobs\endcsname{$+$16000.0}
\expandafter\def\csname val@rev.desc.d9.uiv.natural.zoomclick-d3.haiku.n-requests-match-scores\endcsname{1}
\expandafter\def\csname cnt@rev.desc.d9.uiv.natural.zoomclick-d3.haiku.n-requests-match-scores\endcsname{1}
\expandafter\def\csname pct@rev.desc.d9.uiv.natural.zoomclick-d3.haiku.n-requests-match-scores\endcsname{100.0}
\expandafter\def\csname ppt@rev.desc.d9.uiv.natural.zoomclick-d3.haiku.n-requests-match-scores\endcsname{$+$100.0}
\expandafter\def\csname val@rev.desc.d9.uiv.natural.zoomclick-d3.haiku.ratio-to-direct\endcsname{2.358}
\expandafter\def\csname cnt@rev.desc.d9.uiv.natural.zoomclick-d3.haiku.ratio-to-direct\endcsname{2}
\expandafter\def\csname pct@rev.desc.d9.uiv.natural.zoomclick-d3.haiku.ratio-to-direct\endcsname{235.8}
\expandafter\def\csname ppt@rev.desc.d9.uiv.natural.zoomclick-d3.haiku.ratio-to-direct\endcsname{$+$235.8}
\expandafter\def\csname val@rev.desc.d9.uiv.natural.zoomclick-d3.mean-mpix\endcsname{1.360}
\expandafter\def\csname cnt@rev.desc.d9.uiv.natural.zoomclick-d3.mean-mpix\endcsname{1}
\expandafter\def\csname pct@rev.desc.d9.uiv.natural.zoomclick-d3.mean-mpix\endcsname{136.0}
\expandafter\def\csname ppt@rev.desc.d9.uiv.natural.zoomclick-d3.mean-mpix\endcsname{$+$136.0}
\expandafter\def\csname val@rev.desc.d9.uiv.natural.zoomclick-d3.mean-pixels\endcsname{1359826.816}
\expandafter\def\csname cnt@rev.desc.d9.uiv.natural.zoomclick-d3.mean-pixels\endcsname{1{,}359{,}827}
\expandafter\def\csname pct@rev.desc.d9.uiv.natural.zoomclick-d3.mean-pixels\endcsname{135982681.6}
\expandafter\def\csname ppt@rev.desc.d9.uiv.natural.zoomclick-d3.mean-pixels\endcsname{$+$135982681.6}
\expandafter\def\csname val@rev.desc.d9.uiv.natural.zoomclick-d3.mean-reader-mpix\endcsname{0.351}
\expandafter\def\csname cnt@rev.desc.d9.uiv.natural.zoomclick-d3.mean-reader-mpix\endcsname{0}
\expandafter\def\csname pct@rev.desc.d9.uiv.natural.zoomclick-d3.mean-reader-mpix\endcsname{35.1}
\expandafter\def\csname ppt@rev.desc.d9.uiv.natural.zoomclick-d3.mean-reader-mpix\endcsname{$+$35.1}
\expandafter\def\csname val@rev.desc.d9.uiv.natural.zoomclick-d3.mean-requests\endcsname{2.995}
\expandafter\def\csname cnt@rev.desc.d9.uiv.natural.zoomclick-d3.mean-requests\endcsname{3}
\expandafter\def\csname pct@rev.desc.d9.uiv.natural.zoomclick-d3.mean-requests\endcsname{299.5}
\expandafter\def\csname ppt@rev.desc.d9.uiv.natural.zoomclick-d3.mean-requests\endcsname{$+$299.5}
\expandafter\def\csname val@rev.desc.d9.uiv.natural.zoomclick-d3.n-jobs\endcsname{768}
\expandafter\def\csname cnt@rev.desc.d9.uiv.natural.zoomclick-d3.n-jobs\endcsname{768}
\expandafter\def\csname pct@rev.desc.d9.uiv.natural.zoomclick-d3.n-jobs\endcsname{76800.0}
\expandafter\def\csname ppt@rev.desc.d9.uiv.natural.zoomclick-d3.n-jobs\endcsname{$+$76800.0}
\expandafter\def\csname val@rev.desc.d9.uiv.natural.zoomclick-d3.n-requests-match-scores\endcsname{1}
\expandafter\def\csname cnt@rev.desc.d9.uiv.natural.zoomclick-d3.n-requests-match-scores\endcsname{1}
\expandafter\def\csname pct@rev.desc.d9.uiv.natural.zoomclick-d3.n-requests-match-scores\endcsname{100.0}
\expandafter\def\csname ppt@rev.desc.d9.uiv.natural.zoomclick-d3.n-requests-match-scores\endcsname{$+$100.0}
\expandafter\def\csname val@rev.desc.d9.uiv.natural.zoomclick-d3.opus.mean-mpix\endcsname{1.354}
\expandafter\def\csname cnt@rev.desc.d9.uiv.natural.zoomclick-d3.opus.mean-mpix\endcsname{1}
\expandafter\def\csname pct@rev.desc.d9.uiv.natural.zoomclick-d3.opus.mean-mpix\endcsname{135.4}
\expandafter\def\csname ppt@rev.desc.d9.uiv.natural.zoomclick-d3.opus.mean-mpix\endcsname{$+$135.4}
\expandafter\def\csname val@rev.desc.d9.uiv.natural.zoomclick-d3.opus.mean-pixels\endcsname{1354471.378}
\expandafter\def\csname cnt@rev.desc.d9.uiv.natural.zoomclick-d3.opus.mean-pixels\endcsname{1{,}354{,}471}
\expandafter\def\csname pct@rev.desc.d9.uiv.natural.zoomclick-d3.opus.mean-pixels\endcsname{135447137.8}
\expandafter\def\csname ppt@rev.desc.d9.uiv.natural.zoomclick-d3.opus.mean-pixels\endcsname{$+$135447137.8}
\expandafter\def\csname val@rev.desc.d9.uiv.natural.zoomclick-d3.opus.mean-reader-mpix\endcsname{0.344}
\expandafter\def\csname cnt@rev.desc.d9.uiv.natural.zoomclick-d3.opus.mean-reader-mpix\endcsname{0}
\expandafter\def\csname pct@rev.desc.d9.uiv.natural.zoomclick-d3.opus.mean-reader-mpix\endcsname{34.4}
\expandafter\def\csname ppt@rev.desc.d9.uiv.natural.zoomclick-d3.opus.mean-reader-mpix\endcsname{$+$34.4}
\expandafter\def\csname val@rev.desc.d9.uiv.natural.zoomclick-d3.opus.mean-requests\endcsname{3.000}
\expandafter\def\csname cnt@rev.desc.d9.uiv.natural.zoomclick-d3.opus.mean-requests\endcsname{3}
\expandafter\def\csname pct@rev.desc.d9.uiv.natural.zoomclick-d3.opus.mean-requests\endcsname{300.0}
\expandafter\def\csname ppt@rev.desc.d9.uiv.natural.zoomclick-d3.opus.mean-requests\endcsname{$+$300.0}
\expandafter\def\csname val@rev.desc.d9.uiv.natural.zoomclick-d3.opus.n-jobs\endcsname{304}
\expandafter\def\csname cnt@rev.desc.d9.uiv.natural.zoomclick-d3.opus.n-jobs\endcsname{304}
\expandafter\def\csname pct@rev.desc.d9.uiv.natural.zoomclick-d3.opus.n-jobs\endcsname{30400.0}
\expandafter\def\csname ppt@rev.desc.d9.uiv.natural.zoomclick-d3.opus.n-jobs\endcsname{$+$30400.0}
\expandafter\def\csname val@rev.desc.d9.uiv.natural.zoomclick-d3.opus.n-requests-match-scores\endcsname{1}
\expandafter\def\csname cnt@rev.desc.d9.uiv.natural.zoomclick-d3.opus.n-requests-match-scores\endcsname{1}
\expandafter\def\csname pct@rev.desc.d9.uiv.natural.zoomclick-d3.opus.n-requests-match-scores\endcsname{100.0}
\expandafter\def\csname ppt@rev.desc.d9.uiv.natural.zoomclick-d3.opus.n-requests-match-scores\endcsname{$+$100.0}
\expandafter\def\csname val@rev.desc.d9.uiv.natural.zoomclick-d3.opus.ratio-to-direct\endcsname{2.308}
\expandafter\def\csname cnt@rev.desc.d9.uiv.natural.zoomclick-d3.opus.ratio-to-direct\endcsname{2}
\expandafter\def\csname pct@rev.desc.d9.uiv.natural.zoomclick-d3.opus.ratio-to-direct\endcsname{230.8}
\expandafter\def\csname ppt@rev.desc.d9.uiv.natural.zoomclick-d3.opus.ratio-to-direct\endcsname{$+$230.8}
\expandafter\def\csname val@rev.desc.d9.uiv.natural.zoomclick-d3.ratio-to-direct\endcsname{2.317}
\expandafter\def\csname cnt@rev.desc.d9.uiv.natural.zoomclick-d3.ratio-to-direct\endcsname{2}
\expandafter\def\csname pct@rev.desc.d9.uiv.natural.zoomclick-d3.ratio-to-direct\endcsname{231.7}
\expandafter\def\csname ppt@rev.desc.d9.uiv.natural.zoomclick-d3.ratio-to-direct\endcsname{$+$231.7}
\expandafter\def\csname val@rev.desc.d9.uiv.natural.zoomclick-d3.sonnet.mean-mpix\endcsname{1.353}
\expandafter\def\csname cnt@rev.desc.d9.uiv.natural.zoomclick-d3.sonnet.mean-mpix\endcsname{1}
\expandafter\def\csname pct@rev.desc.d9.uiv.natural.zoomclick-d3.sonnet.mean-mpix\endcsname{135.3}
\expandafter\def\csname ppt@rev.desc.d9.uiv.natural.zoomclick-d3.sonnet.mean-mpix\endcsname{$+$135.3}
\expandafter\def\csname val@rev.desc.d9.uiv.natural.zoomclick-d3.sonnet.mean-pixels\endcsname{1352683.766}
\expandafter\def\csname cnt@rev.desc.d9.uiv.natural.zoomclick-d3.sonnet.mean-pixels\endcsname{1{,}352{,}684}
\expandafter\def\csname pct@rev.desc.d9.uiv.natural.zoomclick-d3.sonnet.mean-pixels\endcsname{135268376.6}
\expandafter\def\csname ppt@rev.desc.d9.uiv.natural.zoomclick-d3.sonnet.mean-pixels\endcsname{$+$135268376.6}
\expandafter\def\csname val@rev.desc.d9.uiv.natural.zoomclick-d3.sonnet.mean-reader-mpix\endcsname{0.345}
\expandafter\def\csname cnt@rev.desc.d9.uiv.natural.zoomclick-d3.sonnet.mean-reader-mpix\endcsname{0}
\expandafter\def\csname pct@rev.desc.d9.uiv.natural.zoomclick-d3.sonnet.mean-reader-mpix\endcsname{34.5}
\expandafter\def\csname ppt@rev.desc.d9.uiv.natural.zoomclick-d3.sonnet.mean-reader-mpix\endcsname{$+$34.5}
\expandafter\def\csname val@rev.desc.d9.uiv.natural.zoomclick-d3.sonnet.mean-requests\endcsname{3.000}
\expandafter\def\csname cnt@rev.desc.d9.uiv.natural.zoomclick-d3.sonnet.mean-requests\endcsname{3}
\expandafter\def\csname pct@rev.desc.d9.uiv.natural.zoomclick-d3.sonnet.mean-requests\endcsname{300.0}
\expandafter\def\csname ppt@rev.desc.d9.uiv.natural.zoomclick-d3.sonnet.mean-requests\endcsname{$+$300.0}
\expandafter\def\csname val@rev.desc.d9.uiv.natural.zoomclick-d3.sonnet.n-jobs\endcsname{304}
\expandafter\def\csname cnt@rev.desc.d9.uiv.natural.zoomclick-d3.sonnet.n-jobs\endcsname{304}
\expandafter\def\csname pct@rev.desc.d9.uiv.natural.zoomclick-d3.sonnet.n-jobs\endcsname{30400.0}
\expandafter\def\csname ppt@rev.desc.d9.uiv.natural.zoomclick-d3.sonnet.n-jobs\endcsname{$+$30400.0}
\expandafter\def\csname val@rev.desc.d9.uiv.natural.zoomclick-d3.sonnet.n-requests-match-scores\endcsname{1}
\expandafter\def\csname cnt@rev.desc.d9.uiv.natural.zoomclick-d3.sonnet.n-requests-match-scores\endcsname{1}
\expandafter\def\csname pct@rev.desc.d9.uiv.natural.zoomclick-d3.sonnet.n-requests-match-scores\endcsname{100.0}
\expandafter\def\csname ppt@rev.desc.d9.uiv.natural.zoomclick-d3.sonnet.n-requests-match-scores\endcsname{$+$100.0}
\expandafter\def\csname val@rev.desc.d9.uiv.natural.zoomclick-d3.sonnet.ratio-to-direct\endcsname{2.305}
\expandafter\def\csname cnt@rev.desc.d9.uiv.natural.zoomclick-d3.sonnet.ratio-to-direct\endcsname{2}
\expandafter\def\csname pct@rev.desc.d9.uiv.natural.zoomclick-d3.sonnet.ratio-to-direct\endcsname{230.5}
\expandafter\def\csname ppt@rev.desc.d9.uiv.natural.zoomclick-d3.sonnet.ratio-to-direct\endcsname{$+$230.5}
\expandafter\def\csname val@rev.desc.d9.uiv.replay.haiku.mean-mpix\endcsname{0.438}
\expandafter\def\csname cnt@rev.desc.d9.uiv.replay.haiku.mean-mpix\endcsname{0}
\expandafter\def\csname pct@rev.desc.d9.uiv.replay.haiku.mean-mpix\endcsname{43.8}
\expandafter\def\csname ppt@rev.desc.d9.uiv.replay.haiku.mean-mpix\endcsname{$+$43.8}
\expandafter\def\csname val@rev.desc.d9.uiv.replay.haiku.mean-pixels\endcsname{437820.723}
\expandafter\def\csname cnt@rev.desc.d9.uiv.replay.haiku.mean-pixels\endcsname{437{,}821}
\expandafter\def\csname pct@rev.desc.d9.uiv.replay.haiku.mean-pixels\endcsname{43782072.3}
\expandafter\def\csname ppt@rev.desc.d9.uiv.replay.haiku.mean-pixels\endcsname{$+$43782072.3}
\expandafter\def\csname val@rev.desc.d9.uiv.replay.haiku.mean-reader-mpix\endcsname{0.438}
\expandafter\def\csname cnt@rev.desc.d9.uiv.replay.haiku.mean-reader-mpix\endcsname{0}
\expandafter\def\csname pct@rev.desc.d9.uiv.replay.haiku.mean-reader-mpix\endcsname{43.8}
\expandafter\def\csname ppt@rev.desc.d9.uiv.replay.haiku.mean-reader-mpix\endcsname{$+$43.8}
\expandafter\def\csname val@rev.desc.d9.uiv.replay.haiku.mean-requests\endcsname{1.000}
\expandafter\def\csname cnt@rev.desc.d9.uiv.replay.haiku.mean-requests\endcsname{1}
\expandafter\def\csname pct@rev.desc.d9.uiv.replay.haiku.mean-requests\endcsname{100.0}
\expandafter\def\csname ppt@rev.desc.d9.uiv.replay.haiku.mean-requests\endcsname{$+$100.0}
\expandafter\def\csname val@rev.desc.d9.uiv.replay.haiku.n-jobs\endcsname{600}
\expandafter\def\csname cnt@rev.desc.d9.uiv.replay.haiku.n-jobs\endcsname{600}
\expandafter\def\csname pct@rev.desc.d9.uiv.replay.haiku.n-jobs\endcsname{60000.0}
\expandafter\def\csname ppt@rev.desc.d9.uiv.replay.haiku.n-jobs\endcsname{$+$60000.0}
\expandafter\def\csname val@rev.desc.d9.uiv.replay.haiku.n-requests-match-scores\endcsname{1}
\expandafter\def\csname cnt@rev.desc.d9.uiv.replay.haiku.n-requests-match-scores\endcsname{1}
\expandafter\def\csname pct@rev.desc.d9.uiv.replay.haiku.n-requests-match-scores\endcsname{100.0}
\expandafter\def\csname ppt@rev.desc.d9.uiv.replay.haiku.n-requests-match-scores\endcsname{$+$100.0}
\expandafter\def\csname val@rev.desc.d9.uiv.replay.haiku.ratio-to-direct\endcsname{0.746}
\expandafter\def\csname cnt@rev.desc.d9.uiv.replay.haiku.ratio-to-direct\endcsname{1}
\expandafter\def\csname pct@rev.desc.d9.uiv.replay.haiku.ratio-to-direct\endcsname{74.6}
\expandafter\def\csname ppt@rev.desc.d9.uiv.replay.haiku.ratio-to-direct\endcsname{$+$74.6}
\expandafter\def\csname val@rev.desc.d9.uiv.replay.mean-mpix\endcsname{0.434}
\expandafter\def\csname cnt@rev.desc.d9.uiv.replay.mean-mpix\endcsname{0}
\expandafter\def\csname pct@rev.desc.d9.uiv.replay.mean-mpix\endcsname{43.4}
\expandafter\def\csname ppt@rev.desc.d9.uiv.replay.mean-mpix\endcsname{$+$43.4}
\expandafter\def\csname val@rev.desc.d9.uiv.replay.mean-pixels\endcsname{434193.205}
\expandafter\def\csname cnt@rev.desc.d9.uiv.replay.mean-pixels\endcsname{434{,}193}
\expandafter\def\csname pct@rev.desc.d9.uiv.replay.mean-pixels\endcsname{43419320.5}
\expandafter\def\csname ppt@rev.desc.d9.uiv.replay.mean-pixels\endcsname{$+$43419320.5}
\expandafter\def\csname val@rev.desc.d9.uiv.replay.mean-reader-mpix\endcsname{0.434}
\expandafter\def\csname cnt@rev.desc.d9.uiv.replay.mean-reader-mpix\endcsname{0}
\expandafter\def\csname pct@rev.desc.d9.uiv.replay.mean-reader-mpix\endcsname{43.4}
\expandafter\def\csname ppt@rev.desc.d9.uiv.replay.mean-reader-mpix\endcsname{$+$43.4}
\expandafter\def\csname val@rev.desc.d9.uiv.replay.mean-requests\endcsname{1.000}
\expandafter\def\csname cnt@rev.desc.d9.uiv.replay.mean-requests\endcsname{1}
\expandafter\def\csname pct@rev.desc.d9.uiv.replay.mean-requests\endcsname{100.0}
\expandafter\def\csname ppt@rev.desc.d9.uiv.replay.mean-requests\endcsname{$+$100.0}
\expandafter\def\csname val@rev.desc.d9.uiv.replay.n-jobs\endcsname{3520}
\expandafter\def\csname cnt@rev.desc.d9.uiv.replay.n-jobs\endcsname{3{,}520}
\expandafter\def\csname pct@rev.desc.d9.uiv.replay.n-jobs\endcsname{352000.0}
\expandafter\def\csname ppt@rev.desc.d9.uiv.replay.n-jobs\endcsname{$+$352000.0}
\expandafter\def\csname val@rev.desc.d9.uiv.replay.n-requests-match-scores\endcsname{1}
\expandafter\def\csname cnt@rev.desc.d9.uiv.replay.n-requests-match-scores\endcsname{1}
\expandafter\def\csname pct@rev.desc.d9.uiv.replay.n-requests-match-scores\endcsname{100.0}
\expandafter\def\csname ppt@rev.desc.d9.uiv.replay.n-requests-match-scores\endcsname{$+$100.0}
\expandafter\def\csname val@rev.desc.d9.uiv.replay.opus.mean-mpix\endcsname{0.435}
\expandafter\def\csname cnt@rev.desc.d9.uiv.replay.opus.mean-mpix\endcsname{0}
\expandafter\def\csname pct@rev.desc.d9.uiv.replay.opus.mean-mpix\endcsname{43.5}
\expandafter\def\csname ppt@rev.desc.d9.uiv.replay.opus.mean-mpix\endcsname{$+$43.5}
\expandafter\def\csname val@rev.desc.d9.uiv.replay.opus.mean-pixels\endcsname{434882.938}
\expandafter\def\csname cnt@rev.desc.d9.uiv.replay.opus.mean-pixels\endcsname{434{,}883}
\expandafter\def\csname pct@rev.desc.d9.uiv.replay.opus.mean-pixels\endcsname{43488293.8}
\expandafter\def\csname ppt@rev.desc.d9.uiv.replay.opus.mean-pixels\endcsname{$+$43488293.8}
\expandafter\def\csname val@rev.desc.d9.uiv.replay.opus.mean-reader-mpix\endcsname{0.435}
\expandafter\def\csname cnt@rev.desc.d9.uiv.replay.opus.mean-reader-mpix\endcsname{0}
\expandafter\def\csname pct@rev.desc.d9.uiv.replay.opus.mean-reader-mpix\endcsname{43.5}
\expandafter\def\csname ppt@rev.desc.d9.uiv.replay.opus.mean-reader-mpix\endcsname{$+$43.5}
\expandafter\def\csname val@rev.desc.d9.uiv.replay.opus.mean-requests\endcsname{1.000}
\expandafter\def\csname cnt@rev.desc.d9.uiv.replay.opus.mean-requests\endcsname{1}
\expandafter\def\csname pct@rev.desc.d9.uiv.replay.opus.mean-requests\endcsname{100.0}
\expandafter\def\csname ppt@rev.desc.d9.uiv.replay.opus.mean-requests\endcsname{$+$100.0}
\expandafter\def\csname val@rev.desc.d9.uiv.replay.opus.n-jobs\endcsname{1520}
\expandafter\def\csname cnt@rev.desc.d9.uiv.replay.opus.n-jobs\endcsname{1{,}520}
\expandafter\def\csname pct@rev.desc.d9.uiv.replay.opus.n-jobs\endcsname{152000.0}
\expandafter\def\csname ppt@rev.desc.d9.uiv.replay.opus.n-jobs\endcsname{$+$152000.0}
\expandafter\def\csname val@rev.desc.d9.uiv.replay.opus.n-requests-match-scores\endcsname{1}
\expandafter\def\csname cnt@rev.desc.d9.uiv.replay.opus.n-requests-match-scores\endcsname{1}
\expandafter\def\csname pct@rev.desc.d9.uiv.replay.opus.n-requests-match-scores\endcsname{100.0}
\expandafter\def\csname ppt@rev.desc.d9.uiv.replay.opus.n-requests-match-scores\endcsname{$+$100.0}
\expandafter\def\csname val@rev.desc.d9.uiv.replay.opus.ratio-to-direct\endcsname{0.741}
\expandafter\def\csname cnt@rev.desc.d9.uiv.replay.opus.ratio-to-direct\endcsname{1}
\expandafter\def\csname pct@rev.desc.d9.uiv.replay.opus.ratio-to-direct\endcsname{74.1}
\expandafter\def\csname ppt@rev.desc.d9.uiv.replay.opus.ratio-to-direct\endcsname{$+$74.1}
\expandafter\def\csname val@rev.desc.d9.uiv.replay.ratio-to-direct\endcsname{0.740}
\expandafter\def\csname cnt@rev.desc.d9.uiv.replay.ratio-to-direct\endcsname{1}
\expandafter\def\csname pct@rev.desc.d9.uiv.replay.ratio-to-direct\endcsname{74.0}
\expandafter\def\csname ppt@rev.desc.d9.uiv.replay.ratio-to-direct\endcsname{$+$74.0}
\expandafter\def\csname val@rev.desc.d9.uiv.replay.sonnet.mean-mpix\endcsname{0.432}
\expandafter\def\csname cnt@rev.desc.d9.uiv.replay.sonnet.mean-mpix\endcsname{0}
\expandafter\def\csname pct@rev.desc.d9.uiv.replay.sonnet.mean-mpix\endcsname{43.2}
\expandafter\def\csname ppt@rev.desc.d9.uiv.replay.sonnet.mean-mpix\endcsname{$+$43.2}
\expandafter\def\csname val@rev.desc.d9.uiv.replay.sonnet.mean-pixels\endcsname{431889.700}
\expandafter\def\csname cnt@rev.desc.d9.uiv.replay.sonnet.mean-pixels\endcsname{431{,}890}
\expandafter\def\csname pct@rev.desc.d9.uiv.replay.sonnet.mean-pixels\endcsname{43188970.0}
\expandafter\def\csname ppt@rev.desc.d9.uiv.replay.sonnet.mean-pixels\endcsname{$+$43188970.0}
\expandafter\def\csname val@rev.desc.d9.uiv.replay.sonnet.mean-reader-mpix\endcsname{0.432}
\expandafter\def\csname cnt@rev.desc.d9.uiv.replay.sonnet.mean-reader-mpix\endcsname{0}
\expandafter\def\csname pct@rev.desc.d9.uiv.replay.sonnet.mean-reader-mpix\endcsname{43.2}
\expandafter\def\csname ppt@rev.desc.d9.uiv.replay.sonnet.mean-reader-mpix\endcsname{$+$43.2}
\expandafter\def\csname val@rev.desc.d9.uiv.replay.sonnet.mean-requests\endcsname{1.000}
\expandafter\def\csname cnt@rev.desc.d9.uiv.replay.sonnet.mean-requests\endcsname{1}
\expandafter\def\csname pct@rev.desc.d9.uiv.replay.sonnet.mean-requests\endcsname{100.0}
\expandafter\def\csname ppt@rev.desc.d9.uiv.replay.sonnet.mean-requests\endcsname{$+$100.0}
\expandafter\def\csname val@rev.desc.d9.uiv.replay.sonnet.n-jobs\endcsname{1400}
\expandafter\def\csname cnt@rev.desc.d9.uiv.replay.sonnet.n-jobs\endcsname{1{,}400}
\expandafter\def\csname pct@rev.desc.d9.uiv.replay.sonnet.n-jobs\endcsname{140000.0}
\expandafter\def\csname ppt@rev.desc.d9.uiv.replay.sonnet.n-jobs\endcsname{$+$140000.0}
\expandafter\def\csname val@rev.desc.d9.uiv.replay.sonnet.n-requests-match-scores\endcsname{1}
\expandafter\def\csname cnt@rev.desc.d9.uiv.replay.sonnet.n-requests-match-scores\endcsname{1}
\expandafter\def\csname pct@rev.desc.d9.uiv.replay.sonnet.n-requests-match-scores\endcsname{100.0}
\expandafter\def\csname ppt@rev.desc.d9.uiv.replay.sonnet.n-requests-match-scores\endcsname{$+$100.0}
\expandafter\def\csname val@rev.desc.d9.uiv.replay.sonnet.ratio-to-direct\endcsname{0.736}
\expandafter\def\csname cnt@rev.desc.d9.uiv.replay.sonnet.ratio-to-direct\endcsname{1}
\expandafter\def\csname pct@rev.desc.d9.uiv.replay.sonnet.ratio-to-direct\endcsname{73.6}
\expandafter\def\csname ppt@rev.desc.d9.uiv.replay.sonnet.ratio-to-direct\endcsname{$+$73.6}
\expandafter\def\csname val@rev.desc.summary.d1.Tsealed-n\endcsname{20}
\expandafter\def\csname cnt@rev.desc.summary.d1.Tsealed-n\endcsname{20}
\expandafter\def\csname pct@rev.desc.summary.d1.Tsealed-n\endcsname{2000.0}
\expandafter\def\csname ppt@rev.desc.summary.d1.Tsealed-n\endcsname{$+$2000.0}
\expandafter\def\csname val@rev.desc.summary.d1.Tsealed-reproduces\endcsname{20}
\expandafter\def\csname cnt@rev.desc.summary.d1.Tsealed-reproduces\endcsname{20}
\expandafter\def\csname pct@rev.desc.summary.d1.Tsealed-reproduces\endcsname{2000.0}
\expandafter\def\csname ppt@rev.desc.summary.d1.Tsealed-reproduces\endcsname{$+$2000.0}
\expandafter\def\csname val@rev.desc.summary.d1.Tsealed-reproduces-ci\endcsname{20}
\expandafter\def\csname cnt@rev.desc.summary.d1.Tsealed-reproduces-ci\endcsname{20}
\expandafter\def\csname pct@rev.desc.summary.d1.Tsealed-reproduces-ci\endcsname{2000.0}
\expandafter\def\csname ppt@rev.desc.summary.d1.Tsealed-reproduces-ci\endcsname{$+$2000.0}
\expandafter\def\csname val@rev.desc.summary.d1.cross.delta-ci-excludes-0\endcsname{0}
\expandafter\def\csname cnt@rev.desc.summary.d1.cross.delta-ci-excludes-0\endcsname{0}
\expandafter\def\csname pct@rev.desc.summary.d1.cross.delta-ci-excludes-0\endcsname{0.0}
\expandafter\def\csname ppt@rev.desc.summary.d1.cross.delta-ci-excludes-0\endcsname{0.0}
\expandafter\def\csname val@rev.desc.summary.d1.cross.material-flips\endcsname{0}
\expandafter\def\csname cnt@rev.desc.summary.d1.cross.material-flips\endcsname{0}
\expandafter\def\csname pct@rev.desc.summary.d1.cross.material-flips\endcsname{0.0}
\expandafter\def\csname ppt@rev.desc.summary.d1.cross.material-flips\endcsname{0.0}
\expandafter\def\csname val@rev.desc.summary.d1.cross.material-flips-cat\endcsname{0}
\expandafter\def\csname cnt@rev.desc.summary.d1.cross.material-flips-cat\endcsname{0}
\expandafter\def\csname pct@rev.desc.summary.d1.cross.material-flips-cat\endcsname{0.0}
\expandafter\def\csname ppt@rev.desc.summary.d1.cross.material-flips-cat\endcsname{0.0}
\expandafter\def\csname val@rev.desc.summary.d1.cross.n\endcsname{10}
\expandafter\def\csname cnt@rev.desc.summary.d1.cross.n\endcsname{10}
\expandafter\def\csname pct@rev.desc.summary.d1.cross.n\endcsname{1000.0}
\expandafter\def\csname ppt@rev.desc.summary.d1.cross.n\endcsname{$+$1000.0}
\expandafter\def\csname val@rev.desc.summary.d1.cross.n-cat\endcsname{8}
\expandafter\def\csname cnt@rev.desc.summary.d1.cross.n-cat\endcsname{8}
\expandafter\def\csname pct@rev.desc.summary.d1.cross.n-cat\endcsname{800.0}
\expandafter\def\csname ppt@rev.desc.summary.d1.cross.n-cat\endcsname{$+$800.0}
\expandafter\def\csname val@rev.desc.summary.d1.cross.sign-differs\endcsname{0}
\expandafter\def\csname cnt@rev.desc.summary.d1.cross.sign-differs\endcsname{0}
\expandafter\def\csname pct@rev.desc.summary.d1.cross.sign-differs\endcsname{0.0}
\expandafter\def\csname ppt@rev.desc.summary.d1.cross.sign-differs\endcsname{0.0}
\expandafter\def\csname val@rev.desc.summary.d1.cross.sign-differs-cat\endcsname{0}
\expandafter\def\csname cnt@rev.desc.summary.d1.cross.sign-differs-cat\endcsname{0}
\expandafter\def\csname pct@rev.desc.summary.d1.cross.sign-differs-cat\endcsname{0.0}
\expandafter\def\csname ppt@rev.desc.summary.d1.cross.sign-differs-cat\endcsname{0.0}
\expandafter\def\csname val@rev.desc.summary.d1.cross.target-ci-excludes-0\endcsname{10}
\expandafter\def\csname cnt@rev.desc.summary.d1.cross.target-ci-excludes-0\endcsname{10}
\expandafter\def\csname pct@rev.desc.summary.d1.cross.target-ci-excludes-0\endcsname{1000.0}
\expandafter\def\csname ppt@rev.desc.summary.d1.cross.target-ci-excludes-0\endcsname{$+$1000.0}
\expandafter\def\csname val@rev.desc.summary.d1.cross.tost1-equivalent\endcsname{0}
\expandafter\def\csname cnt@rev.desc.summary.d1.cross.tost1-equivalent\endcsname{0}
\expandafter\def\csname pct@rev.desc.summary.d1.cross.tost1-equivalent\endcsname{0.0}
\expandafter\def\csname ppt@rev.desc.summary.d1.cross.tost1-equivalent\endcsname{0.0}
\expandafter\def\csname val@rev.desc.summary.d1.cross.tost5-equivalent\endcsname{0}
\expandafter\def\csname cnt@rev.desc.summary.d1.cross.tost5-equivalent\endcsname{0}
\expandafter\def\csname pct@rev.desc.summary.d1.cross.tost5-equivalent\endcsname{0.0}
\expandafter\def\csname ppt@rev.desc.summary.d1.cross.tost5-equivalent\endcsname{0.0}
\expandafter\def\csname val@rev.desc.summary.d1.cross.zero-estimates\endcsname{0}
\expandafter\def\csname cnt@rev.desc.summary.d1.cross.zero-estimates\endcsname{0}
\expandafter\def\csname pct@rev.desc.summary.d1.cross.zero-estimates\endcsname{0.0}
\expandafter\def\csname ppt@rev.desc.summary.d1.cross.zero-estimates\endcsname{0.0}
\expandafter\def\csname val@rev.desc.summary.d1.same-C.delta-ci-excludes-0\endcsname{2}
\expandafter\def\csname cnt@rev.desc.summary.d1.same-C.delta-ci-excludes-0\endcsname{2}
\expandafter\def\csname pct@rev.desc.summary.d1.same-C.delta-ci-excludes-0\endcsname{200.0}
\expandafter\def\csname ppt@rev.desc.summary.d1.same-C.delta-ci-excludes-0\endcsname{$+$200.0}
\expandafter\def\csname val@rev.desc.summary.d1.same-C.delta-halfwidth.max\endcsname{0.208}
\expandafter\def\csname cnt@rev.desc.summary.d1.same-C.delta-halfwidth.max\endcsname{0}
\expandafter\def\csname pct@rev.desc.summary.d1.same-C.delta-halfwidth.max\endcsname{20.8}
\expandafter\def\csname ppt@rev.desc.summary.d1.same-C.delta-halfwidth.max\endcsname{$+$20.8}
\expandafter\def\csname val@rev.desc.summary.d1.same-C.delta-halfwidth.min\endcsname{0.042}
\expandafter\def\csname cnt@rev.desc.summary.d1.same-C.delta-halfwidth.min\endcsname{0}
\expandafter\def\csname pct@rev.desc.summary.d1.same-C.delta-halfwidth.min\endcsname{4.2}
\expandafter\def\csname ppt@rev.desc.summary.d1.same-C.delta-halfwidth.min\endcsname{$+$4.2}
\expandafter\def\csname val@rev.desc.summary.d1.same-C.delta.max\endcsname{0.115}
\expandafter\def\csname cnt@rev.desc.summary.d1.same-C.delta.max\endcsname{0}
\expandafter\def\csname pct@rev.desc.summary.d1.same-C.delta.max\endcsname{11.5}
\expandafter\def\csname ppt@rev.desc.summary.d1.same-C.delta.max\endcsname{$+$11.5}
\expandafter\def\csname val@rev.desc.summary.d1.same-C.delta.min\endcsname{-0.062}
\expandafter\def\csname cnt@rev.desc.summary.d1.same-C.delta.min\endcsname{0}
\expandafter\def\csname pct@rev.desc.summary.d1.same-C.delta.min\endcsname{$-$6.2}
\expandafter\def\csname ppt@rev.desc.summary.d1.same-C.delta.min\endcsname{$-$6.2}
\expandafter\def\csname val@rev.desc.summary.d1.same-C.material-flips\endcsname{0}
\expandafter\def\csname cnt@rev.desc.summary.d1.same-C.material-flips\endcsname{0}
\expandafter\def\csname pct@rev.desc.summary.d1.same-C.material-flips\endcsname{0.0}
\expandafter\def\csname ppt@rev.desc.summary.d1.same-C.material-flips\endcsname{0.0}
\expandafter\def\csname val@rev.desc.summary.d1.same-C.material-flips-cat\endcsname{0}
\expandafter\def\csname cnt@rev.desc.summary.d1.same-C.material-flips-cat\endcsname{0}
\expandafter\def\csname pct@rev.desc.summary.d1.same-C.material-flips-cat\endcsname{0.0}
\expandafter\def\csname ppt@rev.desc.summary.d1.same-C.material-flips-cat\endcsname{0.0}
\expandafter\def\csname val@rev.desc.summary.d1.same-C.n\endcsname{20}
\expandafter\def\csname cnt@rev.desc.summary.d1.same-C.n\endcsname{20}
\expandafter\def\csname pct@rev.desc.summary.d1.same-C.n\endcsname{2000.0}
\expandafter\def\csname ppt@rev.desc.summary.d1.same-C.n\endcsname{$+$2000.0}
\expandafter\def\csname val@rev.desc.summary.d1.same-C.n-cat\endcsname{16}
\expandafter\def\csname cnt@rev.desc.summary.d1.same-C.n-cat\endcsname{16}
\expandafter\def\csname pct@rev.desc.summary.d1.same-C.n-cat\endcsname{1600.0}
\expandafter\def\csname ppt@rev.desc.summary.d1.same-C.n-cat\endcsname{$+$1600.0}
\expandafter\def\csname val@rev.desc.summary.d1.same-C.sign-differs\endcsname{4}
\expandafter\def\csname cnt@rev.desc.summary.d1.same-C.sign-differs\endcsname{4}
\expandafter\def\csname pct@rev.desc.summary.d1.same-C.sign-differs\endcsname{400.0}
\expandafter\def\csname ppt@rev.desc.summary.d1.same-C.sign-differs\endcsname{$+$400.0}
\expandafter\def\csname val@rev.desc.summary.d1.same-C.sign-differs-cat\endcsname{3}
\expandafter\def\csname cnt@rev.desc.summary.d1.same-C.sign-differs-cat\endcsname{3}
\expandafter\def\csname pct@rev.desc.summary.d1.same-C.sign-differs-cat\endcsname{300.0}
\expandafter\def\csname ppt@rev.desc.summary.d1.same-C.sign-differs-cat\endcsname{$+$300.0}
\expandafter\def\csname val@rev.desc.summary.d1.same-C.target-ci-excludes-0\endcsname{8}
\expandafter\def\csname cnt@rev.desc.summary.d1.same-C.target-ci-excludes-0\endcsname{8}
\expandafter\def\csname pct@rev.desc.summary.d1.same-C.target-ci-excludes-0\endcsname{800.0}
\expandafter\def\csname ppt@rev.desc.summary.d1.same-C.target-ci-excludes-0\endcsname{$+$800.0}
\expandafter\def\csname val@rev.desc.summary.d1.same-C.tost1-equivalent\endcsname{0}
\expandafter\def\csname cnt@rev.desc.summary.d1.same-C.tost1-equivalent\endcsname{0}
\expandafter\def\csname pct@rev.desc.summary.d1.same-C.tost1-equivalent\endcsname{0.0}
\expandafter\def\csname ppt@rev.desc.summary.d1.same-C.tost1-equivalent\endcsname{0.0}
\expandafter\def\csname val@rev.desc.summary.d1.same-C.tost5-equivalent\endcsname{0}
\expandafter\def\csname cnt@rev.desc.summary.d1.same-C.tost5-equivalent\endcsname{0}
\expandafter\def\csname pct@rev.desc.summary.d1.same-C.tost5-equivalent\endcsname{0.0}
\expandafter\def\csname ppt@rev.desc.summary.d1.same-C.tost5-equivalent\endcsname{0.0}
\expandafter\def\csname val@rev.desc.summary.d1.same-C.zero-estimates\endcsname{2}
\expandafter\def\csname cnt@rev.desc.summary.d1.same-C.zero-estimates\endcsname{2}
\expandafter\def\csname pct@rev.desc.summary.d1.same-C.zero-estimates\endcsname{200.0}
\expandafter\def\csname ppt@rev.desc.summary.d1.same-C.zero-estimates\endcsname{$+$200.0}
\expandafter\def\csname val@rev.desc.summary.d1.same-T.delta-ci-excludes-0\endcsname{3}
\expandafter\def\csname cnt@rev.desc.summary.d1.same-T.delta-ci-excludes-0\endcsname{3}
\expandafter\def\csname pct@rev.desc.summary.d1.same-T.delta-ci-excludes-0\endcsname{300.0}
\expandafter\def\csname ppt@rev.desc.summary.d1.same-T.delta-ci-excludes-0\endcsname{$+$300.0}
\expandafter\def\csname val@rev.desc.summary.d1.same-T.delta-halfwidth.max\endcsname{0.104}
\expandafter\def\csname cnt@rev.desc.summary.d1.same-T.delta-halfwidth.max\endcsname{0}
\expandafter\def\csname pct@rev.desc.summary.d1.same-T.delta-halfwidth.max\endcsname{10.4}
\expandafter\def\csname ppt@rev.desc.summary.d1.same-T.delta-halfwidth.max\endcsname{$+$10.4}
\expandafter\def\csname val@rev.desc.summary.d1.same-T.delta-halfwidth.min\endcsname{0.042}
\expandafter\def\csname cnt@rev.desc.summary.d1.same-T.delta-halfwidth.min\endcsname{0}
\expandafter\def\csname pct@rev.desc.summary.d1.same-T.delta-halfwidth.min\endcsname{4.2}
\expandafter\def\csname ppt@rev.desc.summary.d1.same-T.delta-halfwidth.min\endcsname{$+$4.2}
\expandafter\def\csname val@rev.desc.summary.d1.same-T.delta.max\endcsname{0.069}
\expandafter\def\csname cnt@rev.desc.summary.d1.same-T.delta.max\endcsname{0}
\expandafter\def\csname pct@rev.desc.summary.d1.same-T.delta.max\endcsname{6.9}
\expandafter\def\csname ppt@rev.desc.summary.d1.same-T.delta.max\endcsname{$+$6.9}
\expandafter\def\csname val@rev.desc.summary.d1.same-T.delta.min\endcsname{-0.010}
\expandafter\def\csname cnt@rev.desc.summary.d1.same-T.delta.min\endcsname{0}
\expandafter\def\csname pct@rev.desc.summary.d1.same-T.delta.min\endcsname{$-$1.0}
\expandafter\def\csname ppt@rev.desc.summary.d1.same-T.delta.min\endcsname{$-$1.0}
\expandafter\def\csname val@rev.desc.summary.d1.same-T.material-flips\endcsname{0}
\expandafter\def\csname cnt@rev.desc.summary.d1.same-T.material-flips\endcsname{0}
\expandafter\def\csname pct@rev.desc.summary.d1.same-T.material-flips\endcsname{0.0}
\expandafter\def\csname ppt@rev.desc.summary.d1.same-T.material-flips\endcsname{0.0}
\expandafter\def\csname val@rev.desc.summary.d1.same-T.material-flips-cat\endcsname{0}
\expandafter\def\csname cnt@rev.desc.summary.d1.same-T.material-flips-cat\endcsname{0}
\expandafter\def\csname pct@rev.desc.summary.d1.same-T.material-flips-cat\endcsname{0.0}
\expandafter\def\csname ppt@rev.desc.summary.d1.same-T.material-flips-cat\endcsname{0.0}
\expandafter\def\csname val@rev.desc.summary.d1.same-T.n\endcsname{10}
\expandafter\def\csname cnt@rev.desc.summary.d1.same-T.n\endcsname{10}
\expandafter\def\csname pct@rev.desc.summary.d1.same-T.n\endcsname{1000.0}
\expandafter\def\csname ppt@rev.desc.summary.d1.same-T.n\endcsname{$+$1000.0}
\expandafter\def\csname val@rev.desc.summary.d1.same-T.n-cat\endcsname{8}
\expandafter\def\csname cnt@rev.desc.summary.d1.same-T.n-cat\endcsname{8}
\expandafter\def\csname pct@rev.desc.summary.d1.same-T.n-cat\endcsname{800.0}
\expandafter\def\csname ppt@rev.desc.summary.d1.same-T.n-cat\endcsname{$+$800.0}
\expandafter\def\csname val@rev.desc.summary.d1.same-T.sign-differs\endcsname{0}
\expandafter\def\csname cnt@rev.desc.summary.d1.same-T.sign-differs\endcsname{0}
\expandafter\def\csname pct@rev.desc.summary.d1.same-T.sign-differs\endcsname{0.0}
\expandafter\def\csname ppt@rev.desc.summary.d1.same-T.sign-differs\endcsname{0.0}
\expandafter\def\csname val@rev.desc.summary.d1.same-T.sign-differs-cat\endcsname{0}
\expandafter\def\csname cnt@rev.desc.summary.d1.same-T.sign-differs-cat\endcsname{0}
\expandafter\def\csname pct@rev.desc.summary.d1.same-T.sign-differs-cat\endcsname{0.0}
\expandafter\def\csname ppt@rev.desc.summary.d1.same-T.sign-differs-cat\endcsname{0.0}
\expandafter\def\csname val@rev.desc.summary.d1.same-T.target-ci-excludes-0\endcsname{6}
\expandafter\def\csname cnt@rev.desc.summary.d1.same-T.target-ci-excludes-0\endcsname{6}
\expandafter\def\csname pct@rev.desc.summary.d1.same-T.target-ci-excludes-0\endcsname{600.0}
\expandafter\def\csname ppt@rev.desc.summary.d1.same-T.target-ci-excludes-0\endcsname{$+$600.0}
\expandafter\def\csname val@rev.desc.summary.d1.same-T.tost1-equivalent\endcsname{0}
\expandafter\def\csname cnt@rev.desc.summary.d1.same-T.tost1-equivalent\endcsname{0}
\expandafter\def\csname pct@rev.desc.summary.d1.same-T.tost1-equivalent\endcsname{0.0}
\expandafter\def\csname ppt@rev.desc.summary.d1.same-T.tost1-equivalent\endcsname{0.0}
\expandafter\def\csname val@rev.desc.summary.d1.same-T.tost5-equivalent\endcsname{0}
\expandafter\def\csname cnt@rev.desc.summary.d1.same-T.tost5-equivalent\endcsname{0}
\expandafter\def\csname pct@rev.desc.summary.d1.same-T.tost5-equivalent\endcsname{0.0}
\expandafter\def\csname ppt@rev.desc.summary.d1.same-T.tost5-equivalent\endcsname{0.0}
\expandafter\def\csname val@rev.desc.summary.d1.same-T.zero-estimates\endcsname{1}
\expandafter\def\csname cnt@rev.desc.summary.d1.same-T.zero-estimates\endcsname{1}
\expandafter\def\csname pct@rev.desc.summary.d1.same-T.zero-estimates\endcsname{100.0}
\expandafter\def\csname ppt@rev.desc.summary.d1.same-T.zero-estimates\endcsname{$+$100.0}
\expandafter\def\csname val@rev.desc.summary.d2.all.D.rate-any.max\endcsname{0.260}
\expandafter\def\csname cnt@rev.desc.summary.d2.all.D.rate-any.max\endcsname{0}
\expandafter\def\csname pct@rev.desc.summary.d2.all.D.rate-any.max\endcsname{26.0}
\expandafter\def\csname ppt@rev.desc.summary.d2.all.D.rate-any.max\endcsname{$+$26.0}
\expandafter\def\csname val@rev.desc.summary.d2.all.D.rate-any.min\endcsname{0.076}
\expandafter\def\csname cnt@rev.desc.summary.d2.all.D.rate-any.min\endcsname{0}
\expandafter\def\csname pct@rev.desc.summary.d2.all.D.rate-any.min\endcsname{7.6}
\expandafter\def\csname ppt@rev.desc.summary.d2.all.D.rate-any.min\endcsname{$+$7.6}
\expandafter\def\csname val@rev.desc.summary.d2.all.F.rate-any.max\endcsname{0.281}
\expandafter\def\csname cnt@rev.desc.summary.d2.all.F.rate-any.max\endcsname{0}
\expandafter\def\csname pct@rev.desc.summary.d2.all.F.rate-any.max\endcsname{28.1}
\expandafter\def\csname ppt@rev.desc.summary.d2.all.F.rate-any.max\endcsname{$+$28.1}
\expandafter\def\csname val@rev.desc.summary.d2.all.F.rate-any.min\endcsname{0.021}
\expandafter\def\csname cnt@rev.desc.summary.d2.all.F.rate-any.min\endcsname{0}
\expandafter\def\csname pct@rev.desc.summary.d2.all.F.rate-any.min\endcsname{2.1}
\expandafter\def\csname ppt@rev.desc.summary.d2.all.F.rate-any.min\endcsname{$+$2.1}
\expandafter\def\csname val@rev.desc.summary.d2.rep.T.rate-any.max\endcsname{0.104}
\expandafter\def\csname cnt@rev.desc.summary.d2.rep.T.rate-any.max\endcsname{0}
\expandafter\def\csname pct@rev.desc.summary.d2.rep.T.rate-any.max\endcsname{10.4}
\expandafter\def\csname ppt@rev.desc.summary.d2.rep.T.rate-any.max\endcsname{$+$10.4}
\expandafter\def\csname val@rev.desc.summary.d2.rep.T.rate-any.min\endcsname{0.021}
\expandafter\def\csname cnt@rev.desc.summary.d2.rep.T.rate-any.min\endcsname{0}
\expandafter\def\csname pct@rev.desc.summary.d2.rep.T.rate-any.min\endcsname{2.1}
\expandafter\def\csname ppt@rev.desc.summary.d2.rep.T.rate-any.min\endcsname{$+$2.1}
\expandafter\def\csname val@rev.desc.summary.d2.sspro.p1.C.F.rate-any.max\endcsname{0.167}
\expandafter\def\csname cnt@rev.desc.summary.d2.sspro.p1.C.F.rate-any.max\endcsname{0}
\expandafter\def\csname pct@rev.desc.summary.d2.sspro.p1.C.F.rate-any.max\endcsname{16.7}
\expandafter\def\csname ppt@rev.desc.summary.d2.sspro.p1.C.F.rate-any.max\endcsname{$+$16.7}
\expandafter\def\csname val@rev.desc.summary.d2.sspro.p1.C.F.rate-any.min\endcsname{0.104}
\expandafter\def\csname cnt@rev.desc.summary.d2.sspro.p1.C.F.rate-any.min\endcsname{0}
\expandafter\def\csname pct@rev.desc.summary.d2.sspro.p1.C.F.rate-any.min\endcsname{10.4}
\expandafter\def\csname ppt@rev.desc.summary.d2.sspro.p1.C.F.rate-any.min\endcsname{$+$10.4}
\expandafter\def\csname val@rev.desc.summary.d2.sspro.p1.C.F.rate01.max\endcsname{0.125}
\expandafter\def\csname cnt@rev.desc.summary.d2.sspro.p1.C.F.rate01.max\endcsname{0}
\expandafter\def\csname pct@rev.desc.summary.d2.sspro.p1.C.F.rate01.max\endcsname{12.5}
\expandafter\def\csname ppt@rev.desc.summary.d2.sspro.p1.C.F.rate01.max\endcsname{$+$12.5}
\expandafter\def\csname val@rev.desc.summary.d2.sspro.p1.C.F.rate01.min\endcsname{0.062}
\expandafter\def\csname cnt@rev.desc.summary.d2.sspro.p1.C.F.rate01.min\endcsname{0}
\expandafter\def\csname pct@rev.desc.summary.d2.sspro.p1.C.F.rate01.min\endcsname{6.2}
\expandafter\def\csname ppt@rev.desc.summary.d2.sspro.p1.C.F.rate01.min\endcsname{$+$6.2}
\expandafter\def\csname val@rev.desc.summary.d2.sspro.p1.C.F.rate10.max\endcsname{0.042}
\expandafter\def\csname cnt@rev.desc.summary.d2.sspro.p1.C.F.rate10.max\endcsname{0}
\expandafter\def\csname pct@rev.desc.summary.d2.sspro.p1.C.F.rate10.max\endcsname{4.2}
\expandafter\def\csname ppt@rev.desc.summary.d2.sspro.p1.C.F.rate10.max\endcsname{$+$4.2}
\expandafter\def\csname val@rev.desc.summary.d2.sspro.p1.C.F.rate10.min\endcsname{0.021}
\expandafter\def\csname cnt@rev.desc.summary.d2.sspro.p1.C.F.rate10.min\endcsname{0}
\expandafter\def\csname pct@rev.desc.summary.d2.sspro.p1.C.F.rate10.min\endcsname{2.1}
\expandafter\def\csname ppt@rev.desc.summary.d2.sspro.p1.C.F.rate10.min\endcsname{$+$2.1}
\expandafter\def\csname val@rev.desc.summary.d2.sspro.p2.C.F.rate-any.max\endcsname{0.229}
\expandafter\def\csname cnt@rev.desc.summary.d2.sspro.p2.C.F.rate-any.max\endcsname{0}
\expandafter\def\csname pct@rev.desc.summary.d2.sspro.p2.C.F.rate-any.max\endcsname{22.9}
\expandafter\def\csname ppt@rev.desc.summary.d2.sspro.p2.C.F.rate-any.max\endcsname{$+$22.9}
\expandafter\def\csname val@rev.desc.summary.d2.sspro.p2.C.F.rate-any.min\endcsname{0.021}
\expandafter\def\csname cnt@rev.desc.summary.d2.sspro.p2.C.F.rate-any.min\endcsname{0}
\expandafter\def\csname pct@rev.desc.summary.d2.sspro.p2.C.F.rate-any.min\endcsname{2.1}
\expandafter\def\csname ppt@rev.desc.summary.d2.sspro.p2.C.F.rate-any.min\endcsname{$+$2.1}
\expandafter\def\csname val@rev.desc.summary.d2.sspro.p2.C.F.rate01.max\endcsname{0.125}
\expandafter\def\csname cnt@rev.desc.summary.d2.sspro.p2.C.F.rate01.max\endcsname{0}
\expandafter\def\csname pct@rev.desc.summary.d2.sspro.p2.C.F.rate01.max\endcsname{12.5}
\expandafter\def\csname ppt@rev.desc.summary.d2.sspro.p2.C.F.rate01.max\endcsname{$+$12.5}
\expandafter\def\csname val@rev.desc.summary.d2.sspro.p2.C.F.rate01.min\endcsname{0.021}
\expandafter\def\csname cnt@rev.desc.summary.d2.sspro.p2.C.F.rate01.min\endcsname{0}
\expandafter\def\csname pct@rev.desc.summary.d2.sspro.p2.C.F.rate01.min\endcsname{2.1}
\expandafter\def\csname ppt@rev.desc.summary.d2.sspro.p2.C.F.rate01.min\endcsname{$+$2.1}
\expandafter\def\csname val@rev.desc.summary.d2.sspro.p2.C.F.rate10.max\endcsname{0.104}
\expandafter\def\csname cnt@rev.desc.summary.d2.sspro.p2.C.F.rate10.max\endcsname{0}
\expandafter\def\csname pct@rev.desc.summary.d2.sspro.p2.C.F.rate10.max\endcsname{10.4}
\expandafter\def\csname ppt@rev.desc.summary.d2.sspro.p2.C.F.rate10.max\endcsname{$+$10.4}
\expandafter\def\csname val@rev.desc.summary.d2.sspro.p2.C.F.rate10.min\endcsname{0.000}
\expandafter\def\csname cnt@rev.desc.summary.d2.sspro.p2.C.F.rate10.min\endcsname{0}
\expandafter\def\csname pct@rev.desc.summary.d2.sspro.p2.C.F.rate10.min\endcsname{0.0}
\expandafter\def\csname ppt@rev.desc.summary.d2.sspro.p2.C.F.rate10.min\endcsname{0.0}
\expandafter\def\csname val@rev.desc.summary.d2.sspro.p2.T.F.rate-any.max\endcsname{0.135}
\expandafter\def\csname cnt@rev.desc.summary.d2.sspro.p2.T.F.rate-any.max\endcsname{0}
\expandafter\def\csname pct@rev.desc.summary.d2.sspro.p2.T.F.rate-any.max\endcsname{13.5}
\expandafter\def\csname ppt@rev.desc.summary.d2.sspro.p2.T.F.rate-any.max\endcsname{$+$13.5}
\expandafter\def\csname val@rev.desc.summary.d2.sspro.p2.T.F.rate-any.min\endcsname{0.073}
\expandafter\def\csname cnt@rev.desc.summary.d2.sspro.p2.T.F.rate-any.min\endcsname{0}
\expandafter\def\csname pct@rev.desc.summary.d2.sspro.p2.T.F.rate-any.min\endcsname{7.3}
\expandafter\def\csname ppt@rev.desc.summary.d2.sspro.p2.T.F.rate-any.min\endcsname{$+$7.3}
\expandafter\def\csname val@rev.desc.summary.d2.sspro.p2.T.F.rate01.max\endcsname{0.094}
\expandafter\def\csname cnt@rev.desc.summary.d2.sspro.p2.T.F.rate01.max\endcsname{0}
\expandafter\def\csname pct@rev.desc.summary.d2.sspro.p2.T.F.rate01.max\endcsname{9.4}
\expandafter\def\csname ppt@rev.desc.summary.d2.sspro.p2.T.F.rate01.max\endcsname{$+$9.4}
\expandafter\def\csname val@rev.desc.summary.d2.sspro.p2.T.F.rate01.min\endcsname{0.052}
\expandafter\def\csname cnt@rev.desc.summary.d2.sspro.p2.T.F.rate01.min\endcsname{0}
\expandafter\def\csname pct@rev.desc.summary.d2.sspro.p2.T.F.rate01.min\endcsname{5.2}
\expandafter\def\csname ppt@rev.desc.summary.d2.sspro.p2.T.F.rate01.min\endcsname{$+$5.2}
\expandafter\def\csname val@rev.desc.summary.d2.sspro.p2.T.F.rate10.max\endcsname{0.042}
\expandafter\def\csname cnt@rev.desc.summary.d2.sspro.p2.T.F.rate10.max\endcsname{0}
\expandafter\def\csname pct@rev.desc.summary.d2.sspro.p2.T.F.rate10.max\endcsname{4.2}
\expandafter\def\csname ppt@rev.desc.summary.d2.sspro.p2.T.F.rate10.max\endcsname{$+$4.2}
\expandafter\def\csname val@rev.desc.summary.d2.sspro.p2.T.F.rate10.min\endcsname{0.010}
\expandafter\def\csname cnt@rev.desc.summary.d2.sspro.p2.T.F.rate10.min\endcsname{0}
\expandafter\def\csname pct@rev.desc.summary.d2.sspro.p2.T.F.rate10.min\endcsname{1.0}
\expandafter\def\csname ppt@rev.desc.summary.d2.sspro.p2.T.F.rate10.min\endcsname{$+$1.0}
\expandafter\def\csname val@rev.desc.summary.d2.uiv.p1.C.F.rate-any.max\endcsname{0.281}
\expandafter\def\csname cnt@rev.desc.summary.d2.uiv.p1.C.F.rate-any.max\endcsname{0}
\expandafter\def\csname pct@rev.desc.summary.d2.uiv.p1.C.F.rate-any.max\endcsname{28.1}
\expandafter\def\csname ppt@rev.desc.summary.d2.uiv.p1.C.F.rate-any.max\endcsname{$+$28.1}
\expandafter\def\csname val@rev.desc.summary.d2.uiv.p1.C.F.rate-any.min\endcsname{0.219}
\expandafter\def\csname cnt@rev.desc.summary.d2.uiv.p1.C.F.rate-any.min\endcsname{0}
\expandafter\def\csname pct@rev.desc.summary.d2.uiv.p1.C.F.rate-any.min\endcsname{21.9}
\expandafter\def\csname ppt@rev.desc.summary.d2.uiv.p1.C.F.rate-any.min\endcsname{$+$21.9}
\expandafter\def\csname val@rev.desc.summary.d2.uiv.p1.C.F.rate01.max\endcsname{0.260}
\expandafter\def\csname cnt@rev.desc.summary.d2.uiv.p1.C.F.rate01.max\endcsname{0}
\expandafter\def\csname pct@rev.desc.summary.d2.uiv.p1.C.F.rate01.max\endcsname{26.0}
\expandafter\def\csname ppt@rev.desc.summary.d2.uiv.p1.C.F.rate01.max\endcsname{$+$26.0}
\expandafter\def\csname val@rev.desc.summary.d2.uiv.p1.C.F.rate01.min\endcsname{0.198}
\expandafter\def\csname cnt@rev.desc.summary.d2.uiv.p1.C.F.rate01.min\endcsname{0}
\expandafter\def\csname pct@rev.desc.summary.d2.uiv.p1.C.F.rate01.min\endcsname{19.8}
\expandafter\def\csname ppt@rev.desc.summary.d2.uiv.p1.C.F.rate01.min\endcsname{$+$19.8}
\expandafter\def\csname val@rev.desc.summary.d2.uiv.p1.C.F.rate10.max\endcsname{0.021}
\expandafter\def\csname cnt@rev.desc.summary.d2.uiv.p1.C.F.rate10.max\endcsname{0}
\expandafter\def\csname pct@rev.desc.summary.d2.uiv.p1.C.F.rate10.max\endcsname{2.1}
\expandafter\def\csname ppt@rev.desc.summary.d2.uiv.p1.C.F.rate10.max\endcsname{$+$2.1}
\expandafter\def\csname val@rev.desc.summary.d2.uiv.p1.C.F.rate10.min\endcsname{0.010}
\expandafter\def\csname cnt@rev.desc.summary.d2.uiv.p1.C.F.rate10.min\endcsname{0}
\expandafter\def\csname pct@rev.desc.summary.d2.uiv.p1.C.F.rate10.min\endcsname{1.0}
\expandafter\def\csname ppt@rev.desc.summary.d2.uiv.p1.C.F.rate10.min\endcsname{$+$1.0}
\expandafter\def\csname val@rev.desc.summary.d2.uiv.p2.C.F.rate-any.max\endcsname{0.156}
\expandafter\def\csname cnt@rev.desc.summary.d2.uiv.p2.C.F.rate-any.max\endcsname{0}
\expandafter\def\csname pct@rev.desc.summary.d2.uiv.p2.C.F.rate-any.max\endcsname{15.6}
\expandafter\def\csname ppt@rev.desc.summary.d2.uiv.p2.C.F.rate-any.max\endcsname{$+$15.6}
\expandafter\def\csname val@rev.desc.summary.d2.uiv.p2.C.F.rate-any.min\endcsname{0.083}
\expandafter\def\csname cnt@rev.desc.summary.d2.uiv.p2.C.F.rate-any.min\endcsname{0}
\expandafter\def\csname pct@rev.desc.summary.d2.uiv.p2.C.F.rate-any.min\endcsname{8.3}
\expandafter\def\csname ppt@rev.desc.summary.d2.uiv.p2.C.F.rate-any.min\endcsname{$+$8.3}
\expandafter\def\csname val@rev.desc.summary.d2.uiv.p2.C.F.rate01.max\endcsname{0.115}
\expandafter\def\csname cnt@rev.desc.summary.d2.uiv.p2.C.F.rate01.max\endcsname{0}
\expandafter\def\csname pct@rev.desc.summary.d2.uiv.p2.C.F.rate01.max\endcsname{11.5}
\expandafter\def\csname ppt@rev.desc.summary.d2.uiv.p2.C.F.rate01.max\endcsname{$+$11.5}
\expandafter\def\csname val@rev.desc.summary.d2.uiv.p2.C.F.rate01.min\endcsname{0.052}
\expandafter\def\csname cnt@rev.desc.summary.d2.uiv.p2.C.F.rate01.min\endcsname{0}
\expandafter\def\csname pct@rev.desc.summary.d2.uiv.p2.C.F.rate01.min\endcsname{5.2}
\expandafter\def\csname ppt@rev.desc.summary.d2.uiv.p2.C.F.rate01.min\endcsname{$+$5.2}
\expandafter\def\csname val@rev.desc.summary.d2.uiv.p2.C.F.rate10.max\endcsname{0.042}
\expandafter\def\csname cnt@rev.desc.summary.d2.uiv.p2.C.F.rate10.max\endcsname{0}
\expandafter\def\csname pct@rev.desc.summary.d2.uiv.p2.C.F.rate10.max\endcsname{4.2}
\expandafter\def\csname ppt@rev.desc.summary.d2.uiv.p2.C.F.rate10.max\endcsname{$+$4.2}
\expandafter\def\csname val@rev.desc.summary.d2.uiv.p2.C.F.rate10.min\endcsname{0.021}
\expandafter\def\csname cnt@rev.desc.summary.d2.uiv.p2.C.F.rate10.min\endcsname{0}
\expandafter\def\csname pct@rev.desc.summary.d2.uiv.p2.C.F.rate10.min\endcsname{2.1}
\expandafter\def\csname ppt@rev.desc.summary.d2.uiv.p2.C.F.rate10.min\endcsname{$+$2.1}
\expandafter\def\csname val@rev.desc.summary.d2.uiv.p2.T.F.rate-any.max\endcsname{0.174}
\expandafter\def\csname cnt@rev.desc.summary.d2.uiv.p2.T.F.rate-any.max\endcsname{0}
\expandafter\def\csname pct@rev.desc.summary.d2.uiv.p2.T.F.rate-any.max\endcsname{17.4}
\expandafter\def\csname ppt@rev.desc.summary.d2.uiv.p2.T.F.rate-any.max\endcsname{$+$17.4}
\expandafter\def\csname val@rev.desc.summary.d2.uiv.p2.T.F.rate-any.min\endcsname{0.076}
\expandafter\def\csname cnt@rev.desc.summary.d2.uiv.p2.T.F.rate-any.min\endcsname{0}
\expandafter\def\csname pct@rev.desc.summary.d2.uiv.p2.T.F.rate-any.min\endcsname{7.6}
\expandafter\def\csname ppt@rev.desc.summary.d2.uiv.p2.T.F.rate-any.min\endcsname{$+$7.6}
\expandafter\def\csname val@rev.desc.summary.d2.uiv.p2.T.F.rate01.max\endcsname{0.125}
\expandafter\def\csname cnt@rev.desc.summary.d2.uiv.p2.T.F.rate01.max\endcsname{0}
\expandafter\def\csname pct@rev.desc.summary.d2.uiv.p2.T.F.rate01.max\endcsname{12.5}
\expandafter\def\csname ppt@rev.desc.summary.d2.uiv.p2.T.F.rate01.max\endcsname{$+$12.5}
\expandafter\def\csname val@rev.desc.summary.d2.uiv.p2.T.F.rate01.min\endcsname{0.056}
\expandafter\def\csname cnt@rev.desc.summary.d2.uiv.p2.T.F.rate01.min\endcsname{0}
\expandafter\def\csname pct@rev.desc.summary.d2.uiv.p2.T.F.rate01.min\endcsname{5.6}
\expandafter\def\csname ppt@rev.desc.summary.d2.uiv.p2.T.F.rate01.min\endcsname{$+$5.6}
\expandafter\def\csname val@rev.desc.summary.d2.uiv.p2.T.F.rate10.max\endcsname{0.056}
\expandafter\def\csname cnt@rev.desc.summary.d2.uiv.p2.T.F.rate10.max\endcsname{0}
\expandafter\def\csname pct@rev.desc.summary.d2.uiv.p2.T.F.rate10.max\endcsname{5.6}
\expandafter\def\csname ppt@rev.desc.summary.d2.uiv.p2.T.F.rate10.max\endcsname{$+$5.6}
\expandafter\def\csname val@rev.desc.summary.d2.uiv.p2.T.F.rate10.min\endcsname{0.007}
\expandafter\def\csname cnt@rev.desc.summary.d2.uiv.p2.T.F.rate10.min\endcsname{0}
\expandafter\def\csname pct@rev.desc.summary.d2.uiv.p2.T.F.rate10.min\endcsname{0.7}
\expandafter\def\csname ppt@rev.desc.summary.d2.uiv.p2.T.F.rate10.min\endcsname{$+$0.7}
\expandafter\def\csname val@rev.desc.summary.d3.perskill-ci-excludes-0\endcsname{0}
\expandafter\def\csname cnt@rev.desc.summary.d3.perskill-ci-excludes-0\endcsname{0}
\expandafter\def\csname pct@rev.desc.summary.d3.perskill-ci-excludes-0\endcsname{0.0}
\expandafter\def\csname ppt@rev.desc.summary.d3.perskill-ci-excludes-0\endcsname{0.0}
\expandafter\def\csname val@rev.desc.summary.d3.perskill-ci.hi-max\endcsname{0.182}
\expandafter\def\csname cnt@rev.desc.summary.d3.perskill-ci.hi-max\endcsname{0}
\expandafter\def\csname pct@rev.desc.summary.d3.perskill-ci.hi-max\endcsname{18.2}
\expandafter\def\csname ppt@rev.desc.summary.d3.perskill-ci.hi-max\endcsname{$+$18.2}
\expandafter\def\csname val@rev.desc.summary.d3.perskill-ci.lo-min\endcsname{-0.179}
\expandafter\def\csname cnt@rev.desc.summary.d3.perskill-ci.lo-min\endcsname{0}
\expandafter\def\csname pct@rev.desc.summary.d3.perskill-ci.lo-min\endcsname{$-$17.9}
\expandafter\def\csname ppt@rev.desc.summary.d3.perskill-ci.lo-min\endcsname{$-$17.9}
\expandafter\def\csname val@rev.desc.summary.d3.perskill-n\endcsname{24}
\expandafter\def\csname cnt@rev.desc.summary.d3.perskill-n\endcsname{24}
\expandafter\def\csname pct@rev.desc.summary.d3.perskill-n\endcsname{2400.0}
\expandafter\def\csname ppt@rev.desc.summary.d3.perskill-n\endcsname{$+$2400.0}
\expandafter\def\csname val@rev.desc.summary.d3.perskill-resid.max\endcsname{0.081}
\expandafter\def\csname cnt@rev.desc.summary.d3.perskill-resid.max\endcsname{0}
\expandafter\def\csname pct@rev.desc.summary.d3.perskill-resid.max\endcsname{8.1}
\expandafter\def\csname ppt@rev.desc.summary.d3.perskill-resid.max\endcsname{$+$8.1}
\expandafter\def\csname val@rev.desc.summary.d3.perskill-resid.min\endcsname{-0.057}
\expandafter\def\csname cnt@rev.desc.summary.d3.perskill-resid.min\endcsname{0}
\expandafter\def\csname pct@rev.desc.summary.d3.perskill-resid.min\endcsname{$-$5.7}
\expandafter\def\csname ppt@rev.desc.summary.d3.perskill-resid.min\endcsname{$-$5.7}
\expandafter\def\csname val@rev.desc.summary.d3.pooled-ci-excludes-0\endcsname{0}
\expandafter\def\csname cnt@rev.desc.summary.d3.pooled-ci-excludes-0\endcsname{0}
\expandafter\def\csname pct@rev.desc.summary.d3.pooled-ci-excludes-0\endcsname{0.0}
\expandafter\def\csname ppt@rev.desc.summary.d3.pooled-ci-excludes-0\endcsname{0.0}
\expandafter\def\csname val@rev.desc.summary.d3.pooled-ci.hi-max\endcsname{0.045}
\expandafter\def\csname cnt@rev.desc.summary.d3.pooled-ci.hi-max\endcsname{0}
\expandafter\def\csname pct@rev.desc.summary.d3.pooled-ci.hi-max\endcsname{4.5}
\expandafter\def\csname ppt@rev.desc.summary.d3.pooled-ci.hi-max\endcsname{$+$4.5}
\expandafter\def\csname val@rev.desc.summary.d3.pooled-ci.lo-min\endcsname{-0.043}
\expandafter\def\csname cnt@rev.desc.summary.d3.pooled-ci.lo-min\endcsname{0}
\expandafter\def\csname pct@rev.desc.summary.d3.pooled-ci.lo-min\endcsname{$-$4.3}
\expandafter\def\csname ppt@rev.desc.summary.d3.pooled-ci.lo-min\endcsname{$-$4.3}
\expandafter\def\csname val@rev.desc.summary.d3.reproduce-n\endcsname{36}
\expandafter\def\csname cnt@rev.desc.summary.d3.reproduce-n\endcsname{36}
\expandafter\def\csname pct@rev.desc.summary.d3.reproduce-n\endcsname{3600.0}
\expandafter\def\csname ppt@rev.desc.summary.d3.reproduce-n\endcsname{$+$3600.0}
\expandafter\def\csname val@rev.desc.summary.d3.reproduced\endcsname{36}
\expandafter\def\csname cnt@rev.desc.summary.d3.reproduced\endcsname{36}
\expandafter\def\csname pct@rev.desc.summary.d3.reproduced\endcsname{3600.0}
\expandafter\def\csname ppt@rev.desc.summary.d3.reproduced\endcsname{$+$3600.0}
\expandafter\def\csname val@rev.desc.summary.d6.box.gap.max\endcsname{0.220}
\expandafter\def\csname cnt@rev.desc.summary.d6.box.gap.max\endcsname{0}
\expandafter\def\csname pct@rev.desc.summary.d6.box.gap.max\endcsname{22.0}
\expandafter\def\csname ppt@rev.desc.summary.d6.box.gap.max\endcsname{$+$22.0}
\expandafter\def\csname val@rev.desc.summary.d6.box.gap.min\endcsname{0.077}
\expandafter\def\csname cnt@rev.desc.summary.d6.box.gap.min\endcsname{0}
\expandafter\def\csname pct@rev.desc.summary.d6.box.gap.min\endcsname{7.7}
\expandafter\def\csname ppt@rev.desc.summary.d6.box.gap.min\endcsname{$+$7.7}
\expandafter\def\csname val@rev.desc.summary.d6.box.gap.n\endcsname{6}
\expandafter\def\csname cnt@rev.desc.summary.d6.box.gap.n\endcsname{6}
\expandafter\def\csname pct@rev.desc.summary.d6.box.gap.n\endcsname{600.0}
\expandafter\def\csname ppt@rev.desc.summary.d6.box.gap.n\endcsname{$+$600.0}
\expandafter\def\csname val@rev.desc.summary.d6.box.gap.n-positive\endcsname{6}
\expandafter\def\csname cnt@rev.desc.summary.d6.box.gap.n-positive\endcsname{6}
\expandafter\def\csname pct@rev.desc.summary.d6.box.gap.n-positive\endcsname{600.0}
\expandafter\def\csname ppt@rev.desc.summary.d6.box.gap.n-positive\endcsname{$+$600.0}
\expandafter\def\csname val@rev.desc.summary.d6.box.perskill-resid.max\endcsname{0.000}
\expandafter\def\csname cnt@rev.desc.summary.d6.box.perskill-resid.max\endcsname{0}
\expandafter\def\csname pct@rev.desc.summary.d6.box.perskill-resid.max\endcsname{0.0}
\expandafter\def\csname ppt@rev.desc.summary.d6.box.perskill-resid.max\endcsname{0.0}
\expandafter\def\csname val@rev.desc.summary.d6.box.perskill-resid.min\endcsname{-0.097}
\expandafter\def\csname cnt@rev.desc.summary.d6.box.perskill-resid.min\endcsname{0}
\expandafter\def\csname pct@rev.desc.summary.d6.box.perskill-resid.min\endcsname{$-$9.7}
\expandafter\def\csname ppt@rev.desc.summary.d6.box.perskill-resid.min\endcsname{$-$9.7}
\expandafter\def\csname val@rev.desc.summary.d6.box.pooled-resid.max\endcsname{-0.004}
\expandafter\def\csname cnt@rev.desc.summary.d6.box.pooled-resid.max\endcsname{0}
\expandafter\def\csname pct@rev.desc.summary.d6.box.pooled-resid.max\endcsname{$-$0.4}
\expandafter\def\csname ppt@rev.desc.summary.d6.box.pooled-resid.max\endcsname{$-$0.4}
\expandafter\def\csname val@rev.desc.summary.d6.box.pooled-resid.min\endcsname{-0.038}
\expandafter\def\csname cnt@rev.desc.summary.d6.box.pooled-resid.min\endcsname{0}
\expandafter\def\csname pct@rev.desc.summary.d6.box.pooled-resid.min\endcsname{$-$3.8}
\expandafter\def\csname ppt@rev.desc.summary.d6.box.pooled-resid.min\endcsname{$-$3.8}
\expandafter\def\csname val@rev.desc.summary.d6.box.share.max\endcsname{0.766}
\expandafter\def\csname cnt@rev.desc.summary.d6.box.share.max\endcsname{1}
\expandafter\def\csname pct@rev.desc.summary.d6.box.share.max\endcsname{76.6}
\expandafter\def\csname ppt@rev.desc.summary.d6.box.share.max\endcsname{$+$76.6}
\expandafter\def\csname val@rev.desc.summary.d6.box.share.min\endcsname{0.568}
\expandafter\def\csname cnt@rev.desc.summary.d6.box.share.min\endcsname{1}
\expandafter\def\csname pct@rev.desc.summary.d6.box.share.min\endcsname{56.8}
\expandafter\def\csname ppt@rev.desc.summary.d6.box.share.min\endcsname{$+$56.8}
\expandafter\def\csname val@rev.desc.summary.d6.tdiag.gap.max\endcsname{0.293}
\expandafter\def\csname cnt@rev.desc.summary.d6.tdiag.gap.max\endcsname{0}
\expandafter\def\csname pct@rev.desc.summary.d6.tdiag.gap.max\endcsname{29.3}
\expandafter\def\csname ppt@rev.desc.summary.d6.tdiag.gap.max\endcsname{$+$29.3}
\expandafter\def\csname val@rev.desc.summary.d6.tdiag.gap.min\endcsname{-0.028}
\expandafter\def\csname cnt@rev.desc.summary.d6.tdiag.gap.min\endcsname{0}
\expandafter\def\csname pct@rev.desc.summary.d6.tdiag.gap.min\endcsname{$-$2.8}
\expandafter\def\csname ppt@rev.desc.summary.d6.tdiag.gap.min\endcsname{$-$2.8}
\expandafter\def\csname val@rev.desc.summary.d6.tdiag.gap.n\endcsname{6}
\expandafter\def\csname cnt@rev.desc.summary.d6.tdiag.gap.n\endcsname{6}
\expandafter\def\csname pct@rev.desc.summary.d6.tdiag.gap.n\endcsname{600.0}
\expandafter\def\csname ppt@rev.desc.summary.d6.tdiag.gap.n\endcsname{$+$600.0}
\expandafter\def\csname val@rev.desc.summary.d6.tdiag.gap.n-positive\endcsname{5}
\expandafter\def\csname cnt@rev.desc.summary.d6.tdiag.gap.n-positive\endcsname{5}
\expandafter\def\csname pct@rev.desc.summary.d6.tdiag.gap.n-positive\endcsname{500.0}
\expandafter\def\csname ppt@rev.desc.summary.d6.tdiag.gap.n-positive\endcsname{$+$500.0}
\expandafter\def\csname val@rev.desc.summary.d6.tdiag.perskill-resid.max\endcsname{0.111}
\expandafter\def\csname cnt@rev.desc.summary.d6.tdiag.perskill-resid.max\endcsname{0}
\expandafter\def\csname pct@rev.desc.summary.d6.tdiag.perskill-resid.max\endcsname{11.1}
\expandafter\def\csname ppt@rev.desc.summary.d6.tdiag.perskill-resid.max\endcsname{$+$11.1}
\expandafter\def\csname val@rev.desc.summary.d6.tdiag.perskill-resid.min\endcsname{-0.083}
\expandafter\def\csname cnt@rev.desc.summary.d6.tdiag.perskill-resid.min\endcsname{0}
\expandafter\def\csname pct@rev.desc.summary.d6.tdiag.perskill-resid.min\endcsname{$-$8.3}
\expandafter\def\csname ppt@rev.desc.summary.d6.tdiag.perskill-resid.min\endcsname{$-$8.3}
\expandafter\def\csname val@rev.desc.summary.d6.tdiag.pooled-resid.max\endcsname{0.028}
\expandafter\def\csname cnt@rev.desc.summary.d6.tdiag.pooled-resid.max\endcsname{0}
\expandafter\def\csname pct@rev.desc.summary.d6.tdiag.pooled-resid.max\endcsname{2.8}
\expandafter\def\csname ppt@rev.desc.summary.d6.tdiag.pooled-resid.max\endcsname{$+$2.8}
\expandafter\def\csname val@rev.desc.summary.d6.tdiag.pooled-resid.min\endcsname{-0.020}
\expandafter\def\csname cnt@rev.desc.summary.d6.tdiag.pooled-resid.min\endcsname{0}
\expandafter\def\csname pct@rev.desc.summary.d6.tdiag.pooled-resid.min\endcsname{$-$2.0}
\expandafter\def\csname ppt@rev.desc.summary.d6.tdiag.pooled-resid.min\endcsname{$-$2.0}
\expandafter\def\csname val@rev.desc.summary.d6.tdiag.share.max\endcsname{0.896}
\expandafter\def\csname cnt@rev.desc.summary.d6.tdiag.share.max\endcsname{1}
\expandafter\def\csname pct@rev.desc.summary.d6.tdiag.share.max\endcsname{89.6}
\expandafter\def\csname ppt@rev.desc.summary.d6.tdiag.share.max\endcsname{$+$89.6}
\expandafter\def\csname val@rev.desc.summary.d6.tdiag.share.min\endcsname{0.719}
\expandafter\def\csname cnt@rev.desc.summary.d6.tdiag.share.min\endcsname{1}
\expandafter\def\csname pct@rev.desc.summary.d6.tdiag.share.min\endcsname{71.9}
\expandafter\def\csname ppt@rev.desc.summary.d6.tdiag.share.min\endcsname{$+$71.9}
\makeatother

\makeatletter
\expandafter\def\csname val@rev.pol.boot-B\endcsname{4000}
\expandafter\def\csname cnt@rev.pol.boot-B\endcsname{4{,}000}
\expandafter\def\csname pct@rev.pol.boot-B\endcsname{400000.0}
\expandafter\def\csname ppt@rev.pol.boot-B\endcsname{$+$400000.0}
\expandafter\def\csname val@rev.pol.boot-B-hi\endcsname{100000}
\expandafter\def\csname cnt@rev.pol.boot-B-hi\endcsname{100{,}000}
\expandafter\def\csname pct@rev.pol.boot-B-hi\endcsname{10000000.0}
\expandafter\def\csname ppt@rev.pol.boot-B-hi\endcsname{$+$10000000.0}
\expandafter\def\csname val@rev.pol.boot.f20.B100000.seed-dependent\endcsname{0}
\expandafter\def\csname cnt@rev.pol.boot.f20.B100000.seed-dependent\endcsname{0}
\expandafter\def\csname pct@rev.pol.boot.f20.B100000.seed-dependent\endcsname{0.0}
\expandafter\def\csname ppt@rev.pol.boot.f20.B100000.seed-dependent\endcsname{0.0}
\expandafter\def\csname val@rev.pol.boot.f20.B100000.sspro.p1.rcr-vs-generic-ppi-source.p-holm-max\endcsname{0.028}
\expandafter\def\csname cnt@rev.pol.boot.f20.B100000.sspro.p1.rcr-vs-generic-ppi-source.p-holm-max\endcsname{0}
\expandafter\def\csname pct@rev.pol.boot.f20.B100000.sspro.p1.rcr-vs-generic-ppi-source.p-holm-max\endcsname{2.8}
\expandafter\def\csname ppt@rev.pol.boot.f20.B100000.sspro.p1.rcr-vs-generic-ppi-source.p-holm-max\endcsname{$+$2.8}
\expandafter\def\csname val@rev.pol.boot.f20.B100000.sspro.p1.rcr-vs-generic-ppi-source.p-holm-min\endcsname{0.027}
\expandafter\def\csname cnt@rev.pol.boot.f20.B100000.sspro.p1.rcr-vs-generic-ppi-source.p-holm-min\endcsname{0}
\expandafter\def\csname pct@rev.pol.boot.f20.B100000.sspro.p1.rcr-vs-generic-ppi-source.p-holm-min\endcsname{2.7}
\expandafter\def\csname ppt@rev.pol.boot.f20.B100000.sspro.p1.rcr-vs-generic-ppi-source.p-holm-min\endcsname{$+$2.7}
\expandafter\def\csname val@rev.pol.boot.f20.B100000.sspro.p1.rcr-vs-replay-ppi-uniform.p-holm-max\endcsname{0.047}
\expandafter\def\csname cnt@rev.pol.boot.f20.B100000.sspro.p1.rcr-vs-replay-ppi-uniform.p-holm-max\endcsname{0}
\expandafter\def\csname pct@rev.pol.boot.f20.B100000.sspro.p1.rcr-vs-replay-ppi-uniform.p-holm-max\endcsname{4.7}
\expandafter\def\csname ppt@rev.pol.boot.f20.B100000.sspro.p1.rcr-vs-replay-ppi-uniform.p-holm-max\endcsname{$+$4.7}
\expandafter\def\csname val@rev.pol.boot.f20.B100000.sspro.p1.rcr-vs-replay-ppi-uniform.p-holm-min\endcsname{0.046}
\expandafter\def\csname cnt@rev.pol.boot.f20.B100000.sspro.p1.rcr-vs-replay-ppi-uniform.p-holm-min\endcsname{0}
\expandafter\def\csname pct@rev.pol.boot.f20.B100000.sspro.p1.rcr-vs-replay-ppi-uniform.p-holm-min\endcsname{4.6}
\expandafter\def\csname ppt@rev.pol.boot.f20.B100000.sspro.p1.rcr-vs-replay-ppi-uniform.p-holm-min\endcsname{$+$4.6}
\expandafter\def\csname val@rev.pol.boot.f20.B100000.sspro.p2.rcr-vs-generic-ppi-source.p-holm-max\endcsname{0.000}
\expandafter\def\csname cnt@rev.pol.boot.f20.B100000.sspro.p2.rcr-vs-generic-ppi-source.p-holm-max\endcsname{0}
\expandafter\def\csname pct@rev.pol.boot.f20.B100000.sspro.p2.rcr-vs-generic-ppi-source.p-holm-max\endcsname{0.0}
\expandafter\def\csname ppt@rev.pol.boot.f20.B100000.sspro.p2.rcr-vs-generic-ppi-source.p-holm-max\endcsname{0.0}
\expandafter\def\csname val@rev.pol.boot.f20.B100000.sspro.p2.rcr-vs-generic-ppi-source.p-holm-min\endcsname{0.000}
\expandafter\def\csname cnt@rev.pol.boot.f20.B100000.sspro.p2.rcr-vs-generic-ppi-source.p-holm-min\endcsname{0}
\expandafter\def\csname pct@rev.pol.boot.f20.B100000.sspro.p2.rcr-vs-generic-ppi-source.p-holm-min\endcsname{0.0}
\expandafter\def\csname ppt@rev.pol.boot.f20.B100000.sspro.p2.rcr-vs-generic-ppi-source.p-holm-min\endcsname{0.0}
\expandafter\def\csname val@rev.pol.boot.f20.B100000.sspro.p2.rcr-vs-replay-ppi-uniform.p-holm-max\endcsname{0.000}
\expandafter\def\csname cnt@rev.pol.boot.f20.B100000.sspro.p2.rcr-vs-replay-ppi-uniform.p-holm-max\endcsname{0}
\expandafter\def\csname pct@rev.pol.boot.f20.B100000.sspro.p2.rcr-vs-replay-ppi-uniform.p-holm-max\endcsname{0.0}
\expandafter\def\csname ppt@rev.pol.boot.f20.B100000.sspro.p2.rcr-vs-replay-ppi-uniform.p-holm-max\endcsname{0.0}
\expandafter\def\csname val@rev.pol.boot.f20.B100000.sspro.p2.rcr-vs-replay-ppi-uniform.p-holm-min\endcsname{0.000}
\expandafter\def\csname cnt@rev.pol.boot.f20.B100000.sspro.p2.rcr-vs-replay-ppi-uniform.p-holm-min\endcsname{0}
\expandafter\def\csname pct@rev.pol.boot.f20.B100000.sspro.p2.rcr-vs-replay-ppi-uniform.p-holm-min\endcsname{0.0}
\expandafter\def\csname ppt@rev.pol.boot.f20.B100000.sspro.p2.rcr-vs-replay-ppi-uniform.p-holm-min\endcsname{0.0}
\expandafter\def\csname val@rev.pol.boot.f20.B100000.uiv.p1.rcr-vs-generic-ppi-source.p-holm-max\endcsname{0.000}
\expandafter\def\csname cnt@rev.pol.boot.f20.B100000.uiv.p1.rcr-vs-generic-ppi-source.p-holm-max\endcsname{0}
\expandafter\def\csname pct@rev.pol.boot.f20.B100000.uiv.p1.rcr-vs-generic-ppi-source.p-holm-max\endcsname{0.0}
\expandafter\def\csname ppt@rev.pol.boot.f20.B100000.uiv.p1.rcr-vs-generic-ppi-source.p-holm-max\endcsname{0.0}
\expandafter\def\csname val@rev.pol.boot.f20.B100000.uiv.p1.rcr-vs-generic-ppi-source.p-holm-min\endcsname{0.000}
\expandafter\def\csname cnt@rev.pol.boot.f20.B100000.uiv.p1.rcr-vs-generic-ppi-source.p-holm-min\endcsname{0}
\expandafter\def\csname pct@rev.pol.boot.f20.B100000.uiv.p1.rcr-vs-generic-ppi-source.p-holm-min\endcsname{0.0}
\expandafter\def\csname ppt@rev.pol.boot.f20.B100000.uiv.p1.rcr-vs-generic-ppi-source.p-holm-min\endcsname{0.0}
\expandafter\def\csname val@rev.pol.boot.f20.B100000.uiv.p1.rcr-vs-replay-ppi-uniform.p-holm-max\endcsname{0.165}
\expandafter\def\csname cnt@rev.pol.boot.f20.B100000.uiv.p1.rcr-vs-replay-ppi-uniform.p-holm-max\endcsname{0}
\expandafter\def\csname pct@rev.pol.boot.f20.B100000.uiv.p1.rcr-vs-replay-ppi-uniform.p-holm-max\endcsname{16.5}
\expandafter\def\csname ppt@rev.pol.boot.f20.B100000.uiv.p1.rcr-vs-replay-ppi-uniform.p-holm-max\endcsname{$+$16.5}
\expandafter\def\csname val@rev.pol.boot.f20.B100000.uiv.p1.rcr-vs-replay-ppi-uniform.p-holm-min\endcsname{0.164}
\expandafter\def\csname cnt@rev.pol.boot.f20.B100000.uiv.p1.rcr-vs-replay-ppi-uniform.p-holm-min\endcsname{0}
\expandafter\def\csname pct@rev.pol.boot.f20.B100000.uiv.p1.rcr-vs-replay-ppi-uniform.p-holm-min\endcsname{16.4}
\expandafter\def\csname ppt@rev.pol.boot.f20.B100000.uiv.p1.rcr-vs-replay-ppi-uniform.p-holm-min\endcsname{$+$16.4}
\expandafter\def\csname val@rev.pol.boot.f20.B100000.uiv.p2.rcr-vs-generic-ppi-source.p-holm-max\endcsname{0.047}
\expandafter\def\csname cnt@rev.pol.boot.f20.B100000.uiv.p2.rcr-vs-generic-ppi-source.p-holm-max\endcsname{0}
\expandafter\def\csname pct@rev.pol.boot.f20.B100000.uiv.p2.rcr-vs-generic-ppi-source.p-holm-max\endcsname{4.7}
\expandafter\def\csname ppt@rev.pol.boot.f20.B100000.uiv.p2.rcr-vs-generic-ppi-source.p-holm-max\endcsname{$+$4.7}
\expandafter\def\csname val@rev.pol.boot.f20.B100000.uiv.p2.rcr-vs-generic-ppi-source.p-holm-min\endcsname{0.046}
\expandafter\def\csname cnt@rev.pol.boot.f20.B100000.uiv.p2.rcr-vs-generic-ppi-source.p-holm-min\endcsname{0}
\expandafter\def\csname pct@rev.pol.boot.f20.B100000.uiv.p2.rcr-vs-generic-ppi-source.p-holm-min\endcsname{4.6}
\expandafter\def\csname ppt@rev.pol.boot.f20.B100000.uiv.p2.rcr-vs-generic-ppi-source.p-holm-min\endcsname{$+$4.6}
\expandafter\def\csname val@rev.pol.boot.f20.B100000.uiv.p2.rcr-vs-replay-ppi-uniform.p-holm-max\endcsname{0.047}
\expandafter\def\csname cnt@rev.pol.boot.f20.B100000.uiv.p2.rcr-vs-replay-ppi-uniform.p-holm-max\endcsname{0}
\expandafter\def\csname pct@rev.pol.boot.f20.B100000.uiv.p2.rcr-vs-replay-ppi-uniform.p-holm-max\endcsname{4.7}
\expandafter\def\csname ppt@rev.pol.boot.f20.B100000.uiv.p2.rcr-vs-replay-ppi-uniform.p-holm-max\endcsname{$+$4.7}
\expandafter\def\csname val@rev.pol.boot.f20.B100000.uiv.p2.rcr-vs-replay-ppi-uniform.p-holm-min\endcsname{0.046}
\expandafter\def\csname cnt@rev.pol.boot.f20.B100000.uiv.p2.rcr-vs-replay-ppi-uniform.p-holm-min\endcsname{0}
\expandafter\def\csname pct@rev.pol.boot.f20.B100000.uiv.p2.rcr-vs-replay-ppi-uniform.p-holm-min\endcsname{4.6}
\expandafter\def\csname ppt@rev.pol.boot.f20.B100000.uiv.p2.rcr-vs-replay-ppi-uniform.p-holm-min\endcsname{$+$4.6}
\expandafter\def\csname val@rev.pol.boot.f20.B4000.seed-dependent\endcsname{0}
\expandafter\def\csname cnt@rev.pol.boot.f20.B4000.seed-dependent\endcsname{0}
\expandafter\def\csname pct@rev.pol.boot.f20.B4000.seed-dependent\endcsname{0.0}
\expandafter\def\csname ppt@rev.pol.boot.f20.B4000.seed-dependent\endcsname{0.0}
\expandafter\def\csname val@rev.pol.boot.f20.content.B100000.max-p-holm-rejected\endcsname{0.047}
\expandafter\def\csname cnt@rev.pol.boot.f20.content.B100000.max-p-holm-rejected\endcsname{0}
\expandafter\def\csname pct@rev.pol.boot.f20.content.B100000.max-p-holm-rejected\endcsname{4.7}
\expandafter\def\csname ppt@rev.pol.boot.f20.content.B100000.max-p-holm-rejected\endcsname{$+$4.7}
\expandafter\def\csname val@rev.pol.boot.f20.content.B100000.n-rejected\endcsname{7}
\expandafter\def\csname cnt@rev.pol.boot.f20.content.B100000.n-rejected\endcsname{7}
\expandafter\def\csname pct@rev.pol.boot.f20.content.B100000.n-rejected\endcsname{700.0}
\expandafter\def\csname ppt@rev.pol.boot.f20.content.B100000.n-rejected\endcsname{$+$700.0}
\expandafter\def\csname val@rev.pol.boot.f20.content.B100000.sspro.p1.rcr-vs-generic-ppi-source.p-holm\endcsname{0.027}
\expandafter\def\csname cnt@rev.pol.boot.f20.content.B100000.sspro.p1.rcr-vs-generic-ppi-source.p-holm\endcsname{0}
\expandafter\def\csname pct@rev.pol.boot.f20.content.B100000.sspro.p1.rcr-vs-generic-ppi-source.p-holm\endcsname{2.7}
\expandafter\def\csname ppt@rev.pol.boot.f20.content.B100000.sspro.p1.rcr-vs-generic-ppi-source.p-holm\endcsname{$+$2.7}
\expandafter\def\csname val@rev.pol.boot.f20.content.B100000.sspro.p1.rcr-vs-generic-ppi-source.p-raw\endcsname{0.005}
\expandafter\def\csname cnt@rev.pol.boot.f20.content.B100000.sspro.p1.rcr-vs-generic-ppi-source.p-raw\endcsname{0}
\expandafter\def\csname pct@rev.pol.boot.f20.content.B100000.sspro.p1.rcr-vs-generic-ppi-source.p-raw\endcsname{0.5}
\expandafter\def\csname ppt@rev.pol.boot.f20.content.B100000.sspro.p1.rcr-vs-generic-ppi-source.p-raw\endcsname{$+$0.5}
\expandafter\def\csname val@rev.pol.boot.f20.content.B100000.sspro.p1.rcr-vs-generic-ppi-source.reject\endcsname{1}
\expandafter\def\csname cnt@rev.pol.boot.f20.content.B100000.sspro.p1.rcr-vs-generic-ppi-source.reject\endcsname{1}
\expandafter\def\csname pct@rev.pol.boot.f20.content.B100000.sspro.p1.rcr-vs-generic-ppi-source.reject\endcsname{100.0}
\expandafter\def\csname ppt@rev.pol.boot.f20.content.B100000.sspro.p1.rcr-vs-generic-ppi-source.reject\endcsname{$+$100.0}
\expandafter\def\csname val@rev.pol.boot.f20.content.B100000.sspro.p1.rcr-vs-replay-ppi-uniform.p-holm\endcsname{0.047}
\expandafter\def\csname cnt@rev.pol.boot.f20.content.B100000.sspro.p1.rcr-vs-replay-ppi-uniform.p-holm\endcsname{0}
\expandafter\def\csname pct@rev.pol.boot.f20.content.B100000.sspro.p1.rcr-vs-replay-ppi-uniform.p-holm\endcsname{4.7}
\expandafter\def\csname ppt@rev.pol.boot.f20.content.B100000.sspro.p1.rcr-vs-replay-ppi-uniform.p-holm\endcsname{$+$4.7}
\expandafter\def\csname val@rev.pol.boot.f20.content.B100000.sspro.p1.rcr-vs-replay-ppi-uniform.p-raw\endcsname{0.013}
\expandafter\def\csname cnt@rev.pol.boot.f20.content.B100000.sspro.p1.rcr-vs-replay-ppi-uniform.p-raw\endcsname{0}
\expandafter\def\csname pct@rev.pol.boot.f20.content.B100000.sspro.p1.rcr-vs-replay-ppi-uniform.p-raw\endcsname{1.3}
\expandafter\def\csname ppt@rev.pol.boot.f20.content.B100000.sspro.p1.rcr-vs-replay-ppi-uniform.p-raw\endcsname{$+$1.3}
\expandafter\def\csname val@rev.pol.boot.f20.content.B100000.sspro.p1.rcr-vs-replay-ppi-uniform.reject\endcsname{1}
\expandafter\def\csname cnt@rev.pol.boot.f20.content.B100000.sspro.p1.rcr-vs-replay-ppi-uniform.reject\endcsname{1}
\expandafter\def\csname pct@rev.pol.boot.f20.content.B100000.sspro.p1.rcr-vs-replay-ppi-uniform.reject\endcsname{100.0}
\expandafter\def\csname ppt@rev.pol.boot.f20.content.B100000.sspro.p1.rcr-vs-replay-ppi-uniform.reject\endcsname{$+$100.0}
\expandafter\def\csname val@rev.pol.boot.f20.content.B100000.sspro.p2.rcr-vs-generic-ppi-source.p-holm\endcsname{0.000}
\expandafter\def\csname cnt@rev.pol.boot.f20.content.B100000.sspro.p2.rcr-vs-generic-ppi-source.p-holm\endcsname{0}
\expandafter\def\csname pct@rev.pol.boot.f20.content.B100000.sspro.p2.rcr-vs-generic-ppi-source.p-holm\endcsname{0.0}
\expandafter\def\csname ppt@rev.pol.boot.f20.content.B100000.sspro.p2.rcr-vs-generic-ppi-source.p-holm\endcsname{0.0}
\expandafter\def\csname val@rev.pol.boot.f20.content.B100000.sspro.p2.rcr-vs-generic-ppi-source.p-raw\endcsname{0.000}
\expandafter\def\csname cnt@rev.pol.boot.f20.content.B100000.sspro.p2.rcr-vs-generic-ppi-source.p-raw\endcsname{0}
\expandafter\def\csname pct@rev.pol.boot.f20.content.B100000.sspro.p2.rcr-vs-generic-ppi-source.p-raw\endcsname{0.0}
\expandafter\def\csname ppt@rev.pol.boot.f20.content.B100000.sspro.p2.rcr-vs-generic-ppi-source.p-raw\endcsname{0.0}
\expandafter\def\csname val@rev.pol.boot.f20.content.B100000.sspro.p2.rcr-vs-generic-ppi-source.reject\endcsname{1}
\expandafter\def\csname cnt@rev.pol.boot.f20.content.B100000.sspro.p2.rcr-vs-generic-ppi-source.reject\endcsname{1}
\expandafter\def\csname pct@rev.pol.boot.f20.content.B100000.sspro.p2.rcr-vs-generic-ppi-source.reject\endcsname{100.0}
\expandafter\def\csname ppt@rev.pol.boot.f20.content.B100000.sspro.p2.rcr-vs-generic-ppi-source.reject\endcsname{$+$100.0}
\expandafter\def\csname val@rev.pol.boot.f20.content.B100000.sspro.p2.rcr-vs-replay-ppi-uniform.p-holm\endcsname{0.000}
\expandafter\def\csname cnt@rev.pol.boot.f20.content.B100000.sspro.p2.rcr-vs-replay-ppi-uniform.p-holm\endcsname{0}
\expandafter\def\csname pct@rev.pol.boot.f20.content.B100000.sspro.p2.rcr-vs-replay-ppi-uniform.p-holm\endcsname{0.0}
\expandafter\def\csname ppt@rev.pol.boot.f20.content.B100000.sspro.p2.rcr-vs-replay-ppi-uniform.p-holm\endcsname{0.0}
\expandafter\def\csname val@rev.pol.boot.f20.content.B100000.sspro.p2.rcr-vs-replay-ppi-uniform.p-raw\endcsname{0.000}
\expandafter\def\csname cnt@rev.pol.boot.f20.content.B100000.sspro.p2.rcr-vs-replay-ppi-uniform.p-raw\endcsname{0}
\expandafter\def\csname pct@rev.pol.boot.f20.content.B100000.sspro.p2.rcr-vs-replay-ppi-uniform.p-raw\endcsname{0.0}
\expandafter\def\csname ppt@rev.pol.boot.f20.content.B100000.sspro.p2.rcr-vs-replay-ppi-uniform.p-raw\endcsname{0.0}
\expandafter\def\csname val@rev.pol.boot.f20.content.B100000.sspro.p2.rcr-vs-replay-ppi-uniform.reject\endcsname{1}
\expandafter\def\csname cnt@rev.pol.boot.f20.content.B100000.sspro.p2.rcr-vs-replay-ppi-uniform.reject\endcsname{1}
\expandafter\def\csname pct@rev.pol.boot.f20.content.B100000.sspro.p2.rcr-vs-replay-ppi-uniform.reject\endcsname{100.0}
\expandafter\def\csname ppt@rev.pol.boot.f20.content.B100000.sspro.p2.rcr-vs-replay-ppi-uniform.reject\endcsname{$+$100.0}
\expandafter\def\csname val@rev.pol.boot.f20.content.B100000.uiv.p1.rcr-vs-generic-ppi-source.p-holm\endcsname{0.000}
\expandafter\def\csname cnt@rev.pol.boot.f20.content.B100000.uiv.p1.rcr-vs-generic-ppi-source.p-holm\endcsname{0}
\expandafter\def\csname pct@rev.pol.boot.f20.content.B100000.uiv.p1.rcr-vs-generic-ppi-source.p-holm\endcsname{0.0}
\expandafter\def\csname ppt@rev.pol.boot.f20.content.B100000.uiv.p1.rcr-vs-generic-ppi-source.p-holm\endcsname{0.0}
\expandafter\def\csname val@rev.pol.boot.f20.content.B100000.uiv.p1.rcr-vs-generic-ppi-source.p-raw\endcsname{0.000}
\expandafter\def\csname cnt@rev.pol.boot.f20.content.B100000.uiv.p1.rcr-vs-generic-ppi-source.p-raw\endcsname{0}
\expandafter\def\csname pct@rev.pol.boot.f20.content.B100000.uiv.p1.rcr-vs-generic-ppi-source.p-raw\endcsname{0.0}
\expandafter\def\csname ppt@rev.pol.boot.f20.content.B100000.uiv.p1.rcr-vs-generic-ppi-source.p-raw\endcsname{0.0}
\expandafter\def\csname val@rev.pol.boot.f20.content.B100000.uiv.p1.rcr-vs-generic-ppi-source.reject\endcsname{1}
\expandafter\def\csname cnt@rev.pol.boot.f20.content.B100000.uiv.p1.rcr-vs-generic-ppi-source.reject\endcsname{1}
\expandafter\def\csname pct@rev.pol.boot.f20.content.B100000.uiv.p1.rcr-vs-generic-ppi-source.reject\endcsname{100.0}
\expandafter\def\csname ppt@rev.pol.boot.f20.content.B100000.uiv.p1.rcr-vs-generic-ppi-source.reject\endcsname{$+$100.0}
\expandafter\def\csname val@rev.pol.boot.f20.content.B100000.uiv.p1.rcr-vs-replay-ppi-uniform.p-holm\endcsname{0.164}
\expandafter\def\csname cnt@rev.pol.boot.f20.content.B100000.uiv.p1.rcr-vs-replay-ppi-uniform.p-holm\endcsname{0}
\expandafter\def\csname pct@rev.pol.boot.f20.content.B100000.uiv.p1.rcr-vs-replay-ppi-uniform.p-holm\endcsname{16.4}
\expandafter\def\csname ppt@rev.pol.boot.f20.content.B100000.uiv.p1.rcr-vs-replay-ppi-uniform.p-holm\endcsname{$+$16.4}
\expandafter\def\csname val@rev.pol.boot.f20.content.B100000.uiv.p1.rcr-vs-replay-ppi-uniform.p-raw\endcsname{0.164}
\expandafter\def\csname cnt@rev.pol.boot.f20.content.B100000.uiv.p1.rcr-vs-replay-ppi-uniform.p-raw\endcsname{0}
\expandafter\def\csname pct@rev.pol.boot.f20.content.B100000.uiv.p1.rcr-vs-replay-ppi-uniform.p-raw\endcsname{16.4}
\expandafter\def\csname ppt@rev.pol.boot.f20.content.B100000.uiv.p1.rcr-vs-replay-ppi-uniform.p-raw\endcsname{$+$16.4}
\expandafter\def\csname val@rev.pol.boot.f20.content.B100000.uiv.p1.rcr-vs-replay-ppi-uniform.reject\endcsname{0}
\expandafter\def\csname cnt@rev.pol.boot.f20.content.B100000.uiv.p1.rcr-vs-replay-ppi-uniform.reject\endcsname{0}
\expandafter\def\csname pct@rev.pol.boot.f20.content.B100000.uiv.p1.rcr-vs-replay-ppi-uniform.reject\endcsname{0.0}
\expandafter\def\csname ppt@rev.pol.boot.f20.content.B100000.uiv.p1.rcr-vs-replay-ppi-uniform.reject\endcsname{0.0}
\expandafter\def\csname val@rev.pol.boot.f20.content.B100000.uiv.p2.rcr-vs-generic-ppi-source.p-holm\endcsname{0.047}
\expandafter\def\csname cnt@rev.pol.boot.f20.content.B100000.uiv.p2.rcr-vs-generic-ppi-source.p-holm\endcsname{0}
\expandafter\def\csname pct@rev.pol.boot.f20.content.B100000.uiv.p2.rcr-vs-generic-ppi-source.p-holm\endcsname{4.7}
\expandafter\def\csname ppt@rev.pol.boot.f20.content.B100000.uiv.p2.rcr-vs-generic-ppi-source.p-holm\endcsname{$+$4.7}
\expandafter\def\csname val@rev.pol.boot.f20.content.B100000.uiv.p2.rcr-vs-generic-ppi-source.p-raw\endcsname{0.012}
\expandafter\def\csname cnt@rev.pol.boot.f20.content.B100000.uiv.p2.rcr-vs-generic-ppi-source.p-raw\endcsname{0}
\expandafter\def\csname pct@rev.pol.boot.f20.content.B100000.uiv.p2.rcr-vs-generic-ppi-source.p-raw\endcsname{1.2}
\expandafter\def\csname ppt@rev.pol.boot.f20.content.B100000.uiv.p2.rcr-vs-generic-ppi-source.p-raw\endcsname{$+$1.2}
\expandafter\def\csname val@rev.pol.boot.f20.content.B100000.uiv.p2.rcr-vs-generic-ppi-source.reject\endcsname{1}
\expandafter\def\csname cnt@rev.pol.boot.f20.content.B100000.uiv.p2.rcr-vs-generic-ppi-source.reject\endcsname{1}
\expandafter\def\csname pct@rev.pol.boot.f20.content.B100000.uiv.p2.rcr-vs-generic-ppi-source.reject\endcsname{100.0}
\expandafter\def\csname ppt@rev.pol.boot.f20.content.B100000.uiv.p2.rcr-vs-generic-ppi-source.reject\endcsname{$+$100.0}
\expandafter\def\csname val@rev.pol.boot.f20.content.B100000.uiv.p2.rcr-vs-replay-ppi-uniform.p-holm\endcsname{0.047}
\expandafter\def\csname cnt@rev.pol.boot.f20.content.B100000.uiv.p2.rcr-vs-replay-ppi-uniform.p-holm\endcsname{0}
\expandafter\def\csname pct@rev.pol.boot.f20.content.B100000.uiv.p2.rcr-vs-replay-ppi-uniform.p-holm\endcsname{4.7}
\expandafter\def\csname ppt@rev.pol.boot.f20.content.B100000.uiv.p2.rcr-vs-replay-ppi-uniform.p-holm\endcsname{$+$4.7}
\expandafter\def\csname val@rev.pol.boot.f20.content.B100000.uiv.p2.rcr-vs-replay-ppi-uniform.p-raw\endcsname{0.013}
\expandafter\def\csname cnt@rev.pol.boot.f20.content.B100000.uiv.p2.rcr-vs-replay-ppi-uniform.p-raw\endcsname{0}
\expandafter\def\csname pct@rev.pol.boot.f20.content.B100000.uiv.p2.rcr-vs-replay-ppi-uniform.p-raw\endcsname{1.3}
\expandafter\def\csname ppt@rev.pol.boot.f20.content.B100000.uiv.p2.rcr-vs-replay-ppi-uniform.p-raw\endcsname{$+$1.3}
\expandafter\def\csname val@rev.pol.boot.f20.content.B100000.uiv.p2.rcr-vs-replay-ppi-uniform.reject\endcsname{1}
\expandafter\def\csname cnt@rev.pol.boot.f20.content.B100000.uiv.p2.rcr-vs-replay-ppi-uniform.reject\endcsname{1}
\expandafter\def\csname pct@rev.pol.boot.f20.content.B100000.uiv.p2.rcr-vs-replay-ppi-uniform.reject\endcsname{100.0}
\expandafter\def\csname ppt@rev.pol.boot.f20.content.B100000.uiv.p2.rcr-vs-replay-ppi-uniform.reject\endcsname{$+$100.0}
\expandafter\def\csname val@rev.pol.boot.f20.content.B4000.n-rejected\endcsname{7}
\expandafter\def\csname cnt@rev.pol.boot.f20.content.B4000.n-rejected\endcsname{7}
\expandafter\def\csname pct@rev.pol.boot.f20.content.B4000.n-rejected\endcsname{700.0}
\expandafter\def\csname ppt@rev.pol.boot.f20.content.B4000.n-rejected\endcsname{$+$700.0}
\expandafter\def\csname val@rev.pol.boot.f20.content.B4000.sspro.p1.rcr-vs-generic-ppi-source.p-holm\endcsname{0.021}
\expandafter\def\csname cnt@rev.pol.boot.f20.content.B4000.sspro.p1.rcr-vs-generic-ppi-source.p-holm\endcsname{0}
\expandafter\def\csname pct@rev.pol.boot.f20.content.B4000.sspro.p1.rcr-vs-generic-ppi-source.p-holm\endcsname{2.1}
\expandafter\def\csname ppt@rev.pol.boot.f20.content.B4000.sspro.p1.rcr-vs-generic-ppi-source.p-holm\endcsname{$+$2.1}
\expandafter\def\csname val@rev.pol.boot.f20.content.B4000.sspro.p1.rcr-vs-generic-ppi-source.p-raw\endcsname{0.004}
\expandafter\def\csname cnt@rev.pol.boot.f20.content.B4000.sspro.p1.rcr-vs-generic-ppi-source.p-raw\endcsname{0}
\expandafter\def\csname pct@rev.pol.boot.f20.content.B4000.sspro.p1.rcr-vs-generic-ppi-source.p-raw\endcsname{0.4}
\expandafter\def\csname ppt@rev.pol.boot.f20.content.B4000.sspro.p1.rcr-vs-generic-ppi-source.p-raw\endcsname{$+$0.4}
\expandafter\def\csname val@rev.pol.boot.f20.content.B4000.sspro.p1.rcr-vs-generic-ppi-source.reject\endcsname{1}
\expandafter\def\csname cnt@rev.pol.boot.f20.content.B4000.sspro.p1.rcr-vs-generic-ppi-source.reject\endcsname{1}
\expandafter\def\csname pct@rev.pol.boot.f20.content.B4000.sspro.p1.rcr-vs-generic-ppi-source.reject\endcsname{100.0}
\expandafter\def\csname ppt@rev.pol.boot.f20.content.B4000.sspro.p1.rcr-vs-generic-ppi-source.reject\endcsname{$+$100.0}
\expandafter\def\csname val@rev.pol.boot.f20.content.B4000.sspro.p1.rcr-vs-replay-ppi-uniform.p-holm\endcsname{0.049}
\expandafter\def\csname cnt@rev.pol.boot.f20.content.B4000.sspro.p1.rcr-vs-replay-ppi-uniform.p-holm\endcsname{0}
\expandafter\def\csname pct@rev.pol.boot.f20.content.B4000.sspro.p1.rcr-vs-replay-ppi-uniform.p-holm\endcsname{4.9}
\expandafter\def\csname ppt@rev.pol.boot.f20.content.B4000.sspro.p1.rcr-vs-replay-ppi-uniform.p-holm\endcsname{$+$4.9}
\expandafter\def\csname val@rev.pol.boot.f20.content.B4000.sspro.p1.rcr-vs-replay-ppi-uniform.p-raw\endcsname{0.014}
\expandafter\def\csname cnt@rev.pol.boot.f20.content.B4000.sspro.p1.rcr-vs-replay-ppi-uniform.p-raw\endcsname{0}
\expandafter\def\csname pct@rev.pol.boot.f20.content.B4000.sspro.p1.rcr-vs-replay-ppi-uniform.p-raw\endcsname{1.4}
\expandafter\def\csname ppt@rev.pol.boot.f20.content.B4000.sspro.p1.rcr-vs-replay-ppi-uniform.p-raw\endcsname{$+$1.4}
\expandafter\def\csname val@rev.pol.boot.f20.content.B4000.sspro.p1.rcr-vs-replay-ppi-uniform.reject\endcsname{1}
\expandafter\def\csname cnt@rev.pol.boot.f20.content.B4000.sspro.p1.rcr-vs-replay-ppi-uniform.reject\endcsname{1}
\expandafter\def\csname pct@rev.pol.boot.f20.content.B4000.sspro.p1.rcr-vs-replay-ppi-uniform.reject\endcsname{100.0}
\expandafter\def\csname ppt@rev.pol.boot.f20.content.B4000.sspro.p1.rcr-vs-replay-ppi-uniform.reject\endcsname{$+$100.0}
\expandafter\def\csname val@rev.pol.boot.f20.content.B4000.sspro.p2.rcr-vs-generic-ppi-source.p-holm\endcsname{0.002}
\expandafter\def\csname cnt@rev.pol.boot.f20.content.B4000.sspro.p2.rcr-vs-generic-ppi-source.p-holm\endcsname{0}
\expandafter\def\csname pct@rev.pol.boot.f20.content.B4000.sspro.p2.rcr-vs-generic-ppi-source.p-holm\endcsname{0.2}
\expandafter\def\csname ppt@rev.pol.boot.f20.content.B4000.sspro.p2.rcr-vs-generic-ppi-source.p-holm\endcsname{$+$0.2}
\expandafter\def\csname val@rev.pol.boot.f20.content.B4000.sspro.p2.rcr-vs-generic-ppi-source.p-raw\endcsname{0.000}
\expandafter\def\csname cnt@rev.pol.boot.f20.content.B4000.sspro.p2.rcr-vs-generic-ppi-source.p-raw\endcsname{0}
\expandafter\def\csname pct@rev.pol.boot.f20.content.B4000.sspro.p2.rcr-vs-generic-ppi-source.p-raw\endcsname{0.0}
\expandafter\def\csname ppt@rev.pol.boot.f20.content.B4000.sspro.p2.rcr-vs-generic-ppi-source.p-raw\endcsname{0.0}
\expandafter\def\csname val@rev.pol.boot.f20.content.B4000.sspro.p2.rcr-vs-generic-ppi-source.reject\endcsname{1}
\expandafter\def\csname cnt@rev.pol.boot.f20.content.B4000.sspro.p2.rcr-vs-generic-ppi-source.reject\endcsname{1}
\expandafter\def\csname pct@rev.pol.boot.f20.content.B4000.sspro.p2.rcr-vs-generic-ppi-source.reject\endcsname{100.0}
\expandafter\def\csname ppt@rev.pol.boot.f20.content.B4000.sspro.p2.rcr-vs-generic-ppi-source.reject\endcsname{$+$100.0}
\expandafter\def\csname val@rev.pol.boot.f20.content.B4000.sspro.p2.rcr-vs-replay-ppi-uniform.p-holm\endcsname{0.002}
\expandafter\def\csname cnt@rev.pol.boot.f20.content.B4000.sspro.p2.rcr-vs-replay-ppi-uniform.p-holm\endcsname{0}
\expandafter\def\csname pct@rev.pol.boot.f20.content.B4000.sspro.p2.rcr-vs-replay-ppi-uniform.p-holm\endcsname{0.2}
\expandafter\def\csname ppt@rev.pol.boot.f20.content.B4000.sspro.p2.rcr-vs-replay-ppi-uniform.p-holm\endcsname{$+$0.2}
\expandafter\def\csname val@rev.pol.boot.f20.content.B4000.sspro.p2.rcr-vs-replay-ppi-uniform.p-raw\endcsname{0.000}
\expandafter\def\csname cnt@rev.pol.boot.f20.content.B4000.sspro.p2.rcr-vs-replay-ppi-uniform.p-raw\endcsname{0}
\expandafter\def\csname pct@rev.pol.boot.f20.content.B4000.sspro.p2.rcr-vs-replay-ppi-uniform.p-raw\endcsname{0.0}
\expandafter\def\csname ppt@rev.pol.boot.f20.content.B4000.sspro.p2.rcr-vs-replay-ppi-uniform.p-raw\endcsname{0.0}
\expandafter\def\csname val@rev.pol.boot.f20.content.B4000.sspro.p2.rcr-vs-replay-ppi-uniform.reject\endcsname{1}
\expandafter\def\csname cnt@rev.pol.boot.f20.content.B4000.sspro.p2.rcr-vs-replay-ppi-uniform.reject\endcsname{1}
\expandafter\def\csname pct@rev.pol.boot.f20.content.B4000.sspro.p2.rcr-vs-replay-ppi-uniform.reject\endcsname{100.0}
\expandafter\def\csname ppt@rev.pol.boot.f20.content.B4000.sspro.p2.rcr-vs-replay-ppi-uniform.reject\endcsname{$+$100.0}
\expandafter\def\csname val@rev.pol.boot.f20.content.B4000.uiv.p1.rcr-vs-generic-ppi-source.p-holm\endcsname{0.002}
\expandafter\def\csname cnt@rev.pol.boot.f20.content.B4000.uiv.p1.rcr-vs-generic-ppi-source.p-holm\endcsname{0}
\expandafter\def\csname pct@rev.pol.boot.f20.content.B4000.uiv.p1.rcr-vs-generic-ppi-source.p-holm\endcsname{0.2}
\expandafter\def\csname ppt@rev.pol.boot.f20.content.B4000.uiv.p1.rcr-vs-generic-ppi-source.p-holm\endcsname{$+$0.2}
\expandafter\def\csname val@rev.pol.boot.f20.content.B4000.uiv.p1.rcr-vs-generic-ppi-source.p-raw\endcsname{0.000}
\expandafter\def\csname cnt@rev.pol.boot.f20.content.B4000.uiv.p1.rcr-vs-generic-ppi-source.p-raw\endcsname{0}
\expandafter\def\csname pct@rev.pol.boot.f20.content.B4000.uiv.p1.rcr-vs-generic-ppi-source.p-raw\endcsname{0.0}
\expandafter\def\csname ppt@rev.pol.boot.f20.content.B4000.uiv.p1.rcr-vs-generic-ppi-source.p-raw\endcsname{0.0}
\expandafter\def\csname val@rev.pol.boot.f20.content.B4000.uiv.p1.rcr-vs-generic-ppi-source.reject\endcsname{1}
\expandafter\def\csname cnt@rev.pol.boot.f20.content.B4000.uiv.p1.rcr-vs-generic-ppi-source.reject\endcsname{1}
\expandafter\def\csname pct@rev.pol.boot.f20.content.B4000.uiv.p1.rcr-vs-generic-ppi-source.reject\endcsname{100.0}
\expandafter\def\csname ppt@rev.pol.boot.f20.content.B4000.uiv.p1.rcr-vs-generic-ppi-source.reject\endcsname{$+$100.0}
\expandafter\def\csname val@rev.pol.boot.f20.content.B4000.uiv.p1.rcr-vs-replay-ppi-uniform.p-holm\endcsname{0.165}
\expandafter\def\csname cnt@rev.pol.boot.f20.content.B4000.uiv.p1.rcr-vs-replay-ppi-uniform.p-holm\endcsname{0}
\expandafter\def\csname pct@rev.pol.boot.f20.content.B4000.uiv.p1.rcr-vs-replay-ppi-uniform.p-holm\endcsname{16.5}
\expandafter\def\csname ppt@rev.pol.boot.f20.content.B4000.uiv.p1.rcr-vs-replay-ppi-uniform.p-holm\endcsname{$+$16.5}
\expandafter\def\csname val@rev.pol.boot.f20.content.B4000.uiv.p1.rcr-vs-replay-ppi-uniform.p-raw\endcsname{0.165}
\expandafter\def\csname cnt@rev.pol.boot.f20.content.B4000.uiv.p1.rcr-vs-replay-ppi-uniform.p-raw\endcsname{0}
\expandafter\def\csname pct@rev.pol.boot.f20.content.B4000.uiv.p1.rcr-vs-replay-ppi-uniform.p-raw\endcsname{16.5}
\expandafter\def\csname ppt@rev.pol.boot.f20.content.B4000.uiv.p1.rcr-vs-replay-ppi-uniform.p-raw\endcsname{$+$16.5}
\expandafter\def\csname val@rev.pol.boot.f20.content.B4000.uiv.p1.rcr-vs-replay-ppi-uniform.reject\endcsname{0}
\expandafter\def\csname cnt@rev.pol.boot.f20.content.B4000.uiv.p1.rcr-vs-replay-ppi-uniform.reject\endcsname{0}
\expandafter\def\csname pct@rev.pol.boot.f20.content.B4000.uiv.p1.rcr-vs-replay-ppi-uniform.reject\endcsname{0.0}
\expandafter\def\csname ppt@rev.pol.boot.f20.content.B4000.uiv.p1.rcr-vs-replay-ppi-uniform.reject\endcsname{0.0}
\expandafter\def\csname val@rev.pol.boot.f20.content.B4000.uiv.p2.rcr-vs-generic-ppi-source.p-holm\endcsname{0.049}
\expandafter\def\csname cnt@rev.pol.boot.f20.content.B4000.uiv.p2.rcr-vs-generic-ppi-source.p-holm\endcsname{0}
\expandafter\def\csname pct@rev.pol.boot.f20.content.B4000.uiv.p2.rcr-vs-generic-ppi-source.p-holm\endcsname{4.9}
\expandafter\def\csname ppt@rev.pol.boot.f20.content.B4000.uiv.p2.rcr-vs-generic-ppi-source.p-holm\endcsname{$+$4.9}
\expandafter\def\csname val@rev.pol.boot.f20.content.B4000.uiv.p2.rcr-vs-generic-ppi-source.p-raw\endcsname{0.012}
\expandafter\def\csname cnt@rev.pol.boot.f20.content.B4000.uiv.p2.rcr-vs-generic-ppi-source.p-raw\endcsname{0}
\expandafter\def\csname pct@rev.pol.boot.f20.content.B4000.uiv.p2.rcr-vs-generic-ppi-source.p-raw\endcsname{1.2}
\expandafter\def\csname ppt@rev.pol.boot.f20.content.B4000.uiv.p2.rcr-vs-generic-ppi-source.p-raw\endcsname{$+$1.2}
\expandafter\def\csname val@rev.pol.boot.f20.content.B4000.uiv.p2.rcr-vs-generic-ppi-source.reject\endcsname{1}
\expandafter\def\csname cnt@rev.pol.boot.f20.content.B4000.uiv.p2.rcr-vs-generic-ppi-source.reject\endcsname{1}
\expandafter\def\csname pct@rev.pol.boot.f20.content.B4000.uiv.p2.rcr-vs-generic-ppi-source.reject\endcsname{100.0}
\expandafter\def\csname ppt@rev.pol.boot.f20.content.B4000.uiv.p2.rcr-vs-generic-ppi-source.reject\endcsname{$+$100.0}
\expandafter\def\csname val@rev.pol.boot.f20.content.B4000.uiv.p2.rcr-vs-replay-ppi-uniform.p-holm\endcsname{0.049}
\expandafter\def\csname cnt@rev.pol.boot.f20.content.B4000.uiv.p2.rcr-vs-replay-ppi-uniform.p-holm\endcsname{0}
\expandafter\def\csname pct@rev.pol.boot.f20.content.B4000.uiv.p2.rcr-vs-replay-ppi-uniform.p-holm\endcsname{4.9}
\expandafter\def\csname ppt@rev.pol.boot.f20.content.B4000.uiv.p2.rcr-vs-replay-ppi-uniform.p-holm\endcsname{$+$4.9}
\expandafter\def\csname val@rev.pol.boot.f20.content.B4000.uiv.p2.rcr-vs-replay-ppi-uniform.p-raw\endcsname{0.013}
\expandafter\def\csname cnt@rev.pol.boot.f20.content.B4000.uiv.p2.rcr-vs-replay-ppi-uniform.p-raw\endcsname{0}
\expandafter\def\csname pct@rev.pol.boot.f20.content.B4000.uiv.p2.rcr-vs-replay-ppi-uniform.p-raw\endcsname{1.3}
\expandafter\def\csname ppt@rev.pol.boot.f20.content.B4000.uiv.p2.rcr-vs-replay-ppi-uniform.p-raw\endcsname{$+$1.3}
\expandafter\def\csname val@rev.pol.boot.f20.content.B4000.uiv.p2.rcr-vs-replay-ppi-uniform.reject\endcsname{1}
\expandafter\def\csname cnt@rev.pol.boot.f20.content.B4000.uiv.p2.rcr-vs-replay-ppi-uniform.reject\endcsname{1}
\expandafter\def\csname pct@rev.pol.boot.f20.content.B4000.uiv.p2.rcr-vs-replay-ppi-uniform.reject\endcsname{100.0}
\expandafter\def\csname ppt@rev.pol.boot.f20.content.B4000.uiv.p2.rcr-vs-replay-ppi-uniform.reject\endcsname{$+$100.0}
\expandafter\def\csname val@rev.pol.boot.f20.fixedA.B100000.max-p-holm-rejected\endcsname{0.047}
\expandafter\def\csname cnt@rev.pol.boot.f20.fixedA.B100000.max-p-holm-rejected\endcsname{0}
\expandafter\def\csname pct@rev.pol.boot.f20.fixedA.B100000.max-p-holm-rejected\endcsname{4.7}
\expandafter\def\csname ppt@rev.pol.boot.f20.fixedA.B100000.max-p-holm-rejected\endcsname{$+$4.7}
\expandafter\def\csname val@rev.pol.boot.f20.fixedA.B100000.n-rejected\endcsname{7}
\expandafter\def\csname cnt@rev.pol.boot.f20.fixedA.B100000.n-rejected\endcsname{7}
\expandafter\def\csname pct@rev.pol.boot.f20.fixedA.B100000.n-rejected\endcsname{700.0}
\expandafter\def\csname ppt@rev.pol.boot.f20.fixedA.B100000.n-rejected\endcsname{$+$700.0}
\expandafter\def\csname val@rev.pol.boot.f20.fixedA.B100000.sspro.p1.rcr-vs-generic-ppi-source.p-holm\endcsname{0.027}
\expandafter\def\csname cnt@rev.pol.boot.f20.fixedA.B100000.sspro.p1.rcr-vs-generic-ppi-source.p-holm\endcsname{0}
\expandafter\def\csname pct@rev.pol.boot.f20.fixedA.B100000.sspro.p1.rcr-vs-generic-ppi-source.p-holm\endcsname{2.7}
\expandafter\def\csname ppt@rev.pol.boot.f20.fixedA.B100000.sspro.p1.rcr-vs-generic-ppi-source.p-holm\endcsname{$+$2.7}
\expandafter\def\csname val@rev.pol.boot.f20.fixedA.B100000.sspro.p1.rcr-vs-generic-ppi-source.p-raw\endcsname{0.005}
\expandafter\def\csname cnt@rev.pol.boot.f20.fixedA.B100000.sspro.p1.rcr-vs-generic-ppi-source.p-raw\endcsname{0}
\expandafter\def\csname pct@rev.pol.boot.f20.fixedA.B100000.sspro.p1.rcr-vs-generic-ppi-source.p-raw\endcsname{0.5}
\expandafter\def\csname ppt@rev.pol.boot.f20.fixedA.B100000.sspro.p1.rcr-vs-generic-ppi-source.p-raw\endcsname{$+$0.5}
\expandafter\def\csname val@rev.pol.boot.f20.fixedA.B100000.sspro.p1.rcr-vs-generic-ppi-source.reject\endcsname{1}
\expandafter\def\csname cnt@rev.pol.boot.f20.fixedA.B100000.sspro.p1.rcr-vs-generic-ppi-source.reject\endcsname{1}
\expandafter\def\csname pct@rev.pol.boot.f20.fixedA.B100000.sspro.p1.rcr-vs-generic-ppi-source.reject\endcsname{100.0}
\expandafter\def\csname ppt@rev.pol.boot.f20.fixedA.B100000.sspro.p1.rcr-vs-generic-ppi-source.reject\endcsname{$+$100.0}
\expandafter\def\csname val@rev.pol.boot.f20.fixedA.B100000.sspro.p1.rcr-vs-replay-ppi-uniform.p-holm\endcsname{0.047}
\expandafter\def\csname cnt@rev.pol.boot.f20.fixedA.B100000.sspro.p1.rcr-vs-replay-ppi-uniform.p-holm\endcsname{0}
\expandafter\def\csname pct@rev.pol.boot.f20.fixedA.B100000.sspro.p1.rcr-vs-replay-ppi-uniform.p-holm\endcsname{4.7}
\expandafter\def\csname ppt@rev.pol.boot.f20.fixedA.B100000.sspro.p1.rcr-vs-replay-ppi-uniform.p-holm\endcsname{$+$4.7}
\expandafter\def\csname val@rev.pol.boot.f20.fixedA.B100000.sspro.p1.rcr-vs-replay-ppi-uniform.p-raw\endcsname{0.013}
\expandafter\def\csname cnt@rev.pol.boot.f20.fixedA.B100000.sspro.p1.rcr-vs-replay-ppi-uniform.p-raw\endcsname{0}
\expandafter\def\csname pct@rev.pol.boot.f20.fixedA.B100000.sspro.p1.rcr-vs-replay-ppi-uniform.p-raw\endcsname{1.3}
\expandafter\def\csname ppt@rev.pol.boot.f20.fixedA.B100000.sspro.p1.rcr-vs-replay-ppi-uniform.p-raw\endcsname{$+$1.3}
\expandafter\def\csname val@rev.pol.boot.f20.fixedA.B100000.sspro.p1.rcr-vs-replay-ppi-uniform.reject\endcsname{1}
\expandafter\def\csname cnt@rev.pol.boot.f20.fixedA.B100000.sspro.p1.rcr-vs-replay-ppi-uniform.reject\endcsname{1}
\expandafter\def\csname pct@rev.pol.boot.f20.fixedA.B100000.sspro.p1.rcr-vs-replay-ppi-uniform.reject\endcsname{100.0}
\expandafter\def\csname ppt@rev.pol.boot.f20.fixedA.B100000.sspro.p1.rcr-vs-replay-ppi-uniform.reject\endcsname{$+$100.0}
\expandafter\def\csname val@rev.pol.boot.f20.fixedA.B100000.sspro.p2.rcr-vs-generic-ppi-source.p-holm\endcsname{0.000}
\expandafter\def\csname cnt@rev.pol.boot.f20.fixedA.B100000.sspro.p2.rcr-vs-generic-ppi-source.p-holm\endcsname{0}
\expandafter\def\csname pct@rev.pol.boot.f20.fixedA.B100000.sspro.p2.rcr-vs-generic-ppi-source.p-holm\endcsname{0.0}
\expandafter\def\csname ppt@rev.pol.boot.f20.fixedA.B100000.sspro.p2.rcr-vs-generic-ppi-source.p-holm\endcsname{0.0}
\expandafter\def\csname val@rev.pol.boot.f20.fixedA.B100000.sspro.p2.rcr-vs-generic-ppi-source.p-raw\endcsname{0.000}
\expandafter\def\csname cnt@rev.pol.boot.f20.fixedA.B100000.sspro.p2.rcr-vs-generic-ppi-source.p-raw\endcsname{0}
\expandafter\def\csname pct@rev.pol.boot.f20.fixedA.B100000.sspro.p2.rcr-vs-generic-ppi-source.p-raw\endcsname{0.0}
\expandafter\def\csname ppt@rev.pol.boot.f20.fixedA.B100000.sspro.p2.rcr-vs-generic-ppi-source.p-raw\endcsname{0.0}
\expandafter\def\csname val@rev.pol.boot.f20.fixedA.B100000.sspro.p2.rcr-vs-generic-ppi-source.reject\endcsname{1}
\expandafter\def\csname cnt@rev.pol.boot.f20.fixedA.B100000.sspro.p2.rcr-vs-generic-ppi-source.reject\endcsname{1}
\expandafter\def\csname pct@rev.pol.boot.f20.fixedA.B100000.sspro.p2.rcr-vs-generic-ppi-source.reject\endcsname{100.0}
\expandafter\def\csname ppt@rev.pol.boot.f20.fixedA.B100000.sspro.p2.rcr-vs-generic-ppi-source.reject\endcsname{$+$100.0}
\expandafter\def\csname val@rev.pol.boot.f20.fixedA.B100000.sspro.p2.rcr-vs-replay-ppi-uniform.p-holm\endcsname{0.000}
\expandafter\def\csname cnt@rev.pol.boot.f20.fixedA.B100000.sspro.p2.rcr-vs-replay-ppi-uniform.p-holm\endcsname{0}
\expandafter\def\csname pct@rev.pol.boot.f20.fixedA.B100000.sspro.p2.rcr-vs-replay-ppi-uniform.p-holm\endcsname{0.0}
\expandafter\def\csname ppt@rev.pol.boot.f20.fixedA.B100000.sspro.p2.rcr-vs-replay-ppi-uniform.p-holm\endcsname{0.0}
\expandafter\def\csname val@rev.pol.boot.f20.fixedA.B100000.sspro.p2.rcr-vs-replay-ppi-uniform.p-raw\endcsname{0.000}
\expandafter\def\csname cnt@rev.pol.boot.f20.fixedA.B100000.sspro.p2.rcr-vs-replay-ppi-uniform.p-raw\endcsname{0}
\expandafter\def\csname pct@rev.pol.boot.f20.fixedA.B100000.sspro.p2.rcr-vs-replay-ppi-uniform.p-raw\endcsname{0.0}
\expandafter\def\csname ppt@rev.pol.boot.f20.fixedA.B100000.sspro.p2.rcr-vs-replay-ppi-uniform.p-raw\endcsname{0.0}
\expandafter\def\csname val@rev.pol.boot.f20.fixedA.B100000.sspro.p2.rcr-vs-replay-ppi-uniform.reject\endcsname{1}
\expandafter\def\csname cnt@rev.pol.boot.f20.fixedA.B100000.sspro.p2.rcr-vs-replay-ppi-uniform.reject\endcsname{1}
\expandafter\def\csname pct@rev.pol.boot.f20.fixedA.B100000.sspro.p2.rcr-vs-replay-ppi-uniform.reject\endcsname{100.0}
\expandafter\def\csname ppt@rev.pol.boot.f20.fixedA.B100000.sspro.p2.rcr-vs-replay-ppi-uniform.reject\endcsname{$+$100.0}
\expandafter\def\csname val@rev.pol.boot.f20.fixedA.B100000.uiv.p1.rcr-vs-generic-ppi-source.p-holm\endcsname{0.000}
\expandafter\def\csname cnt@rev.pol.boot.f20.fixedA.B100000.uiv.p1.rcr-vs-generic-ppi-source.p-holm\endcsname{0}
\expandafter\def\csname pct@rev.pol.boot.f20.fixedA.B100000.uiv.p1.rcr-vs-generic-ppi-source.p-holm\endcsname{0.0}
\expandafter\def\csname ppt@rev.pol.boot.f20.fixedA.B100000.uiv.p1.rcr-vs-generic-ppi-source.p-holm\endcsname{0.0}
\expandafter\def\csname val@rev.pol.boot.f20.fixedA.B100000.uiv.p1.rcr-vs-generic-ppi-source.p-raw\endcsname{0.000}
\expandafter\def\csname cnt@rev.pol.boot.f20.fixedA.B100000.uiv.p1.rcr-vs-generic-ppi-source.p-raw\endcsname{0}
\expandafter\def\csname pct@rev.pol.boot.f20.fixedA.B100000.uiv.p1.rcr-vs-generic-ppi-source.p-raw\endcsname{0.0}
\expandafter\def\csname ppt@rev.pol.boot.f20.fixedA.B100000.uiv.p1.rcr-vs-generic-ppi-source.p-raw\endcsname{0.0}
\expandafter\def\csname val@rev.pol.boot.f20.fixedA.B100000.uiv.p1.rcr-vs-generic-ppi-source.reject\endcsname{1}
\expandafter\def\csname cnt@rev.pol.boot.f20.fixedA.B100000.uiv.p1.rcr-vs-generic-ppi-source.reject\endcsname{1}
\expandafter\def\csname pct@rev.pol.boot.f20.fixedA.B100000.uiv.p1.rcr-vs-generic-ppi-source.reject\endcsname{100.0}
\expandafter\def\csname ppt@rev.pol.boot.f20.fixedA.B100000.uiv.p1.rcr-vs-generic-ppi-source.reject\endcsname{$+$100.0}
\expandafter\def\csname val@rev.pol.boot.f20.fixedA.B100000.uiv.p1.rcr-vs-replay-ppi-uniform.p-holm\endcsname{0.165}
\expandafter\def\csname cnt@rev.pol.boot.f20.fixedA.B100000.uiv.p1.rcr-vs-replay-ppi-uniform.p-holm\endcsname{0}
\expandafter\def\csname pct@rev.pol.boot.f20.fixedA.B100000.uiv.p1.rcr-vs-replay-ppi-uniform.p-holm\endcsname{16.5}
\expandafter\def\csname ppt@rev.pol.boot.f20.fixedA.B100000.uiv.p1.rcr-vs-replay-ppi-uniform.p-holm\endcsname{$+$16.5}
\expandafter\def\csname val@rev.pol.boot.f20.fixedA.B100000.uiv.p1.rcr-vs-replay-ppi-uniform.p-raw\endcsname{0.165}
\expandafter\def\csname cnt@rev.pol.boot.f20.fixedA.B100000.uiv.p1.rcr-vs-replay-ppi-uniform.p-raw\endcsname{0}
\expandafter\def\csname pct@rev.pol.boot.f20.fixedA.B100000.uiv.p1.rcr-vs-replay-ppi-uniform.p-raw\endcsname{16.5}
\expandafter\def\csname ppt@rev.pol.boot.f20.fixedA.B100000.uiv.p1.rcr-vs-replay-ppi-uniform.p-raw\endcsname{$+$16.5}
\expandafter\def\csname val@rev.pol.boot.f20.fixedA.B100000.uiv.p1.rcr-vs-replay-ppi-uniform.reject\endcsname{0}
\expandafter\def\csname cnt@rev.pol.boot.f20.fixedA.B100000.uiv.p1.rcr-vs-replay-ppi-uniform.reject\endcsname{0}
\expandafter\def\csname pct@rev.pol.boot.f20.fixedA.B100000.uiv.p1.rcr-vs-replay-ppi-uniform.reject\endcsname{0.0}
\expandafter\def\csname ppt@rev.pol.boot.f20.fixedA.B100000.uiv.p1.rcr-vs-replay-ppi-uniform.reject\endcsname{0.0}
\expandafter\def\csname val@rev.pol.boot.f20.fixedA.B100000.uiv.p2.rcr-vs-generic-ppi-source.p-holm\endcsname{0.047}
\expandafter\def\csname cnt@rev.pol.boot.f20.fixedA.B100000.uiv.p2.rcr-vs-generic-ppi-source.p-holm\endcsname{0}
\expandafter\def\csname pct@rev.pol.boot.f20.fixedA.B100000.uiv.p2.rcr-vs-generic-ppi-source.p-holm\endcsname{4.7}
\expandafter\def\csname ppt@rev.pol.boot.f20.fixedA.B100000.uiv.p2.rcr-vs-generic-ppi-source.p-holm\endcsname{$+$4.7}
\expandafter\def\csname val@rev.pol.boot.f20.fixedA.B100000.uiv.p2.rcr-vs-generic-ppi-source.p-raw\endcsname{0.012}
\expandafter\def\csname cnt@rev.pol.boot.f20.fixedA.B100000.uiv.p2.rcr-vs-generic-ppi-source.p-raw\endcsname{0}
\expandafter\def\csname pct@rev.pol.boot.f20.fixedA.B100000.uiv.p2.rcr-vs-generic-ppi-source.p-raw\endcsname{1.2}
\expandafter\def\csname ppt@rev.pol.boot.f20.fixedA.B100000.uiv.p2.rcr-vs-generic-ppi-source.p-raw\endcsname{$+$1.2}
\expandafter\def\csname val@rev.pol.boot.f20.fixedA.B100000.uiv.p2.rcr-vs-generic-ppi-source.reject\endcsname{1}
\expandafter\def\csname cnt@rev.pol.boot.f20.fixedA.B100000.uiv.p2.rcr-vs-generic-ppi-source.reject\endcsname{1}
\expandafter\def\csname pct@rev.pol.boot.f20.fixedA.B100000.uiv.p2.rcr-vs-generic-ppi-source.reject\endcsname{100.0}
\expandafter\def\csname ppt@rev.pol.boot.f20.fixedA.B100000.uiv.p2.rcr-vs-generic-ppi-source.reject\endcsname{$+$100.0}
\expandafter\def\csname val@rev.pol.boot.f20.fixedA.B100000.uiv.p2.rcr-vs-replay-ppi-uniform.p-holm\endcsname{0.047}
\expandafter\def\csname cnt@rev.pol.boot.f20.fixedA.B100000.uiv.p2.rcr-vs-replay-ppi-uniform.p-holm\endcsname{0}
\expandafter\def\csname pct@rev.pol.boot.f20.fixedA.B100000.uiv.p2.rcr-vs-replay-ppi-uniform.p-holm\endcsname{4.7}
\expandafter\def\csname ppt@rev.pol.boot.f20.fixedA.B100000.uiv.p2.rcr-vs-replay-ppi-uniform.p-holm\endcsname{$+$4.7}
\expandafter\def\csname val@rev.pol.boot.f20.fixedA.B100000.uiv.p2.rcr-vs-replay-ppi-uniform.p-raw\endcsname{0.013}
\expandafter\def\csname cnt@rev.pol.boot.f20.fixedA.B100000.uiv.p2.rcr-vs-replay-ppi-uniform.p-raw\endcsname{0}
\expandafter\def\csname pct@rev.pol.boot.f20.fixedA.B100000.uiv.p2.rcr-vs-replay-ppi-uniform.p-raw\endcsname{1.3}
\expandafter\def\csname ppt@rev.pol.boot.f20.fixedA.B100000.uiv.p2.rcr-vs-replay-ppi-uniform.p-raw\endcsname{$+$1.3}
\expandafter\def\csname val@rev.pol.boot.f20.fixedA.B100000.uiv.p2.rcr-vs-replay-ppi-uniform.reject\endcsname{1}
\expandafter\def\csname cnt@rev.pol.boot.f20.fixedA.B100000.uiv.p2.rcr-vs-replay-ppi-uniform.reject\endcsname{1}
\expandafter\def\csname pct@rev.pol.boot.f20.fixedA.B100000.uiv.p2.rcr-vs-replay-ppi-uniform.reject\endcsname{100.0}
\expandafter\def\csname ppt@rev.pol.boot.f20.fixedA.B100000.uiv.p2.rcr-vs-replay-ppi-uniform.reject\endcsname{$+$100.0}
\expandafter\def\csname val@rev.pol.boot.f20.fixedA.B4000.n-rejected\endcsname{7}
\expandafter\def\csname cnt@rev.pol.boot.f20.fixedA.B4000.n-rejected\endcsname{7}
\expandafter\def\csname pct@rev.pol.boot.f20.fixedA.B4000.n-rejected\endcsname{700.0}
\expandafter\def\csname ppt@rev.pol.boot.f20.fixedA.B4000.n-rejected\endcsname{$+$700.0}
\expandafter\def\csname val@rev.pol.boot.f20.fixedA.B4000.sspro.p1.rcr-vs-generic-ppi-source.p-holm\endcsname{0.029}
\expandafter\def\csname cnt@rev.pol.boot.f20.fixedA.B4000.sspro.p1.rcr-vs-generic-ppi-source.p-holm\endcsname{0}
\expandafter\def\csname pct@rev.pol.boot.f20.fixedA.B4000.sspro.p1.rcr-vs-generic-ppi-source.p-holm\endcsname{2.9}
\expandafter\def\csname ppt@rev.pol.boot.f20.fixedA.B4000.sspro.p1.rcr-vs-generic-ppi-source.p-holm\endcsname{$+$2.9}
\expandafter\def\csname val@rev.pol.boot.f20.fixedA.B4000.sspro.p1.rcr-vs-generic-ppi-source.p-raw\endcsname{0.006}
\expandafter\def\csname cnt@rev.pol.boot.f20.fixedA.B4000.sspro.p1.rcr-vs-generic-ppi-source.p-raw\endcsname{0}
\expandafter\def\csname pct@rev.pol.boot.f20.fixedA.B4000.sspro.p1.rcr-vs-generic-ppi-source.p-raw\endcsname{0.6}
\expandafter\def\csname ppt@rev.pol.boot.f20.fixedA.B4000.sspro.p1.rcr-vs-generic-ppi-source.p-raw\endcsname{$+$0.6}
\expandafter\def\csname val@rev.pol.boot.f20.fixedA.B4000.sspro.p1.rcr-vs-generic-ppi-source.reject\endcsname{1}
\expandafter\def\csname cnt@rev.pol.boot.f20.fixedA.B4000.sspro.p1.rcr-vs-generic-ppi-source.reject\endcsname{1}
\expandafter\def\csname pct@rev.pol.boot.f20.fixedA.B4000.sspro.p1.rcr-vs-generic-ppi-source.reject\endcsname{100.0}
\expandafter\def\csname ppt@rev.pol.boot.f20.fixedA.B4000.sspro.p1.rcr-vs-generic-ppi-source.reject\endcsname{$+$100.0}
\expandafter\def\csname val@rev.pol.boot.f20.fixedA.B4000.sspro.p1.rcr-vs-replay-ppi-uniform.p-holm\endcsname{0.041}
\expandafter\def\csname cnt@rev.pol.boot.f20.fixedA.B4000.sspro.p1.rcr-vs-replay-ppi-uniform.p-holm\endcsname{0}
\expandafter\def\csname pct@rev.pol.boot.f20.fixedA.B4000.sspro.p1.rcr-vs-replay-ppi-uniform.p-holm\endcsname{4.1}
\expandafter\def\csname ppt@rev.pol.boot.f20.fixedA.B4000.sspro.p1.rcr-vs-replay-ppi-uniform.p-holm\endcsname{$+$4.1}
\expandafter\def\csname val@rev.pol.boot.f20.fixedA.B4000.sspro.p1.rcr-vs-replay-ppi-uniform.p-raw\endcsname{0.011}
\expandafter\def\csname cnt@rev.pol.boot.f20.fixedA.B4000.sspro.p1.rcr-vs-replay-ppi-uniform.p-raw\endcsname{0}
\expandafter\def\csname pct@rev.pol.boot.f20.fixedA.B4000.sspro.p1.rcr-vs-replay-ppi-uniform.p-raw\endcsname{1.1}
\expandafter\def\csname ppt@rev.pol.boot.f20.fixedA.B4000.sspro.p1.rcr-vs-replay-ppi-uniform.p-raw\endcsname{$+$1.1}
\expandafter\def\csname val@rev.pol.boot.f20.fixedA.B4000.sspro.p1.rcr-vs-replay-ppi-uniform.reject\endcsname{1}
\expandafter\def\csname cnt@rev.pol.boot.f20.fixedA.B4000.sspro.p1.rcr-vs-replay-ppi-uniform.reject\endcsname{1}
\expandafter\def\csname pct@rev.pol.boot.f20.fixedA.B4000.sspro.p1.rcr-vs-replay-ppi-uniform.reject\endcsname{100.0}
\expandafter\def\csname ppt@rev.pol.boot.f20.fixedA.B4000.sspro.p1.rcr-vs-replay-ppi-uniform.reject\endcsname{$+$100.0}
\expandafter\def\csname val@rev.pol.boot.f20.fixedA.B4000.sspro.p2.rcr-vs-generic-ppi-source.p-holm\endcsname{0.002}
\expandafter\def\csname cnt@rev.pol.boot.f20.fixedA.B4000.sspro.p2.rcr-vs-generic-ppi-source.p-holm\endcsname{0}
\expandafter\def\csname pct@rev.pol.boot.f20.fixedA.B4000.sspro.p2.rcr-vs-generic-ppi-source.p-holm\endcsname{0.2}
\expandafter\def\csname ppt@rev.pol.boot.f20.fixedA.B4000.sspro.p2.rcr-vs-generic-ppi-source.p-holm\endcsname{$+$0.2}
\expandafter\def\csname val@rev.pol.boot.f20.fixedA.B4000.sspro.p2.rcr-vs-generic-ppi-source.p-raw\endcsname{0.000}
\expandafter\def\csname cnt@rev.pol.boot.f20.fixedA.B4000.sspro.p2.rcr-vs-generic-ppi-source.p-raw\endcsname{0}
\expandafter\def\csname pct@rev.pol.boot.f20.fixedA.B4000.sspro.p2.rcr-vs-generic-ppi-source.p-raw\endcsname{0.0}
\expandafter\def\csname ppt@rev.pol.boot.f20.fixedA.B4000.sspro.p2.rcr-vs-generic-ppi-source.p-raw\endcsname{0.0}
\expandafter\def\csname val@rev.pol.boot.f20.fixedA.B4000.sspro.p2.rcr-vs-generic-ppi-source.reject\endcsname{1}
\expandafter\def\csname cnt@rev.pol.boot.f20.fixedA.B4000.sspro.p2.rcr-vs-generic-ppi-source.reject\endcsname{1}
\expandafter\def\csname pct@rev.pol.boot.f20.fixedA.B4000.sspro.p2.rcr-vs-generic-ppi-source.reject\endcsname{100.0}
\expandafter\def\csname ppt@rev.pol.boot.f20.fixedA.B4000.sspro.p2.rcr-vs-generic-ppi-source.reject\endcsname{$+$100.0}
\expandafter\def\csname val@rev.pol.boot.f20.fixedA.B4000.sspro.p2.rcr-vs-replay-ppi-uniform.p-holm\endcsname{0.002}
\expandafter\def\csname cnt@rev.pol.boot.f20.fixedA.B4000.sspro.p2.rcr-vs-replay-ppi-uniform.p-holm\endcsname{0}
\expandafter\def\csname pct@rev.pol.boot.f20.fixedA.B4000.sspro.p2.rcr-vs-replay-ppi-uniform.p-holm\endcsname{0.2}
\expandafter\def\csname ppt@rev.pol.boot.f20.fixedA.B4000.sspro.p2.rcr-vs-replay-ppi-uniform.p-holm\endcsname{$+$0.2}
\expandafter\def\csname val@rev.pol.boot.f20.fixedA.B4000.sspro.p2.rcr-vs-replay-ppi-uniform.p-raw\endcsname{0.000}
\expandafter\def\csname cnt@rev.pol.boot.f20.fixedA.B4000.sspro.p2.rcr-vs-replay-ppi-uniform.p-raw\endcsname{0}
\expandafter\def\csname pct@rev.pol.boot.f20.fixedA.B4000.sspro.p2.rcr-vs-replay-ppi-uniform.p-raw\endcsname{0.0}
\expandafter\def\csname ppt@rev.pol.boot.f20.fixedA.B4000.sspro.p2.rcr-vs-replay-ppi-uniform.p-raw\endcsname{0.0}
\expandafter\def\csname val@rev.pol.boot.f20.fixedA.B4000.sspro.p2.rcr-vs-replay-ppi-uniform.reject\endcsname{1}
\expandafter\def\csname cnt@rev.pol.boot.f20.fixedA.B4000.sspro.p2.rcr-vs-replay-ppi-uniform.reject\endcsname{1}
\expandafter\def\csname pct@rev.pol.boot.f20.fixedA.B4000.sspro.p2.rcr-vs-replay-ppi-uniform.reject\endcsname{100.0}
\expandafter\def\csname ppt@rev.pol.boot.f20.fixedA.B4000.sspro.p2.rcr-vs-replay-ppi-uniform.reject\endcsname{$+$100.0}
\expandafter\def\csname val@rev.pol.boot.f20.fixedA.B4000.uiv.p1.rcr-vs-generic-ppi-source.p-holm\endcsname{0.002}
\expandafter\def\csname cnt@rev.pol.boot.f20.fixedA.B4000.uiv.p1.rcr-vs-generic-ppi-source.p-holm\endcsname{0}
\expandafter\def\csname pct@rev.pol.boot.f20.fixedA.B4000.uiv.p1.rcr-vs-generic-ppi-source.p-holm\endcsname{0.2}
\expandafter\def\csname ppt@rev.pol.boot.f20.fixedA.B4000.uiv.p1.rcr-vs-generic-ppi-source.p-holm\endcsname{$+$0.2}
\expandafter\def\csname val@rev.pol.boot.f20.fixedA.B4000.uiv.p1.rcr-vs-generic-ppi-source.p-raw\endcsname{0.000}
\expandafter\def\csname cnt@rev.pol.boot.f20.fixedA.B4000.uiv.p1.rcr-vs-generic-ppi-source.p-raw\endcsname{0}
\expandafter\def\csname pct@rev.pol.boot.f20.fixedA.B4000.uiv.p1.rcr-vs-generic-ppi-source.p-raw\endcsname{0.0}
\expandafter\def\csname ppt@rev.pol.boot.f20.fixedA.B4000.uiv.p1.rcr-vs-generic-ppi-source.p-raw\endcsname{0.0}
\expandafter\def\csname val@rev.pol.boot.f20.fixedA.B4000.uiv.p1.rcr-vs-generic-ppi-source.reject\endcsname{1}
\expandafter\def\csname cnt@rev.pol.boot.f20.fixedA.B4000.uiv.p1.rcr-vs-generic-ppi-source.reject\endcsname{1}
\expandafter\def\csname pct@rev.pol.boot.f20.fixedA.B4000.uiv.p1.rcr-vs-generic-ppi-source.reject\endcsname{100.0}
\expandafter\def\csname ppt@rev.pol.boot.f20.fixedA.B4000.uiv.p1.rcr-vs-generic-ppi-source.reject\endcsname{$+$100.0}
\expandafter\def\csname val@rev.pol.boot.f20.fixedA.B4000.uiv.p1.rcr-vs-replay-ppi-uniform.p-holm\endcsname{0.171}
\expandafter\def\csname cnt@rev.pol.boot.f20.fixedA.B4000.uiv.p1.rcr-vs-replay-ppi-uniform.p-holm\endcsname{0}
\expandafter\def\csname pct@rev.pol.boot.f20.fixedA.B4000.uiv.p1.rcr-vs-replay-ppi-uniform.p-holm\endcsname{17.1}
\expandafter\def\csname ppt@rev.pol.boot.f20.fixedA.B4000.uiv.p1.rcr-vs-replay-ppi-uniform.p-holm\endcsname{$+$17.1}
\expandafter\def\csname val@rev.pol.boot.f20.fixedA.B4000.uiv.p1.rcr-vs-replay-ppi-uniform.p-raw\endcsname{0.171}
\expandafter\def\csname cnt@rev.pol.boot.f20.fixedA.B4000.uiv.p1.rcr-vs-replay-ppi-uniform.p-raw\endcsname{0}
\expandafter\def\csname pct@rev.pol.boot.f20.fixedA.B4000.uiv.p1.rcr-vs-replay-ppi-uniform.p-raw\endcsname{17.1}
\expandafter\def\csname ppt@rev.pol.boot.f20.fixedA.B4000.uiv.p1.rcr-vs-replay-ppi-uniform.p-raw\endcsname{$+$17.1}
\expandafter\def\csname val@rev.pol.boot.f20.fixedA.B4000.uiv.p1.rcr-vs-replay-ppi-uniform.reject\endcsname{0}
\expandafter\def\csname cnt@rev.pol.boot.f20.fixedA.B4000.uiv.p1.rcr-vs-replay-ppi-uniform.reject\endcsname{0}
\expandafter\def\csname pct@rev.pol.boot.f20.fixedA.B4000.uiv.p1.rcr-vs-replay-ppi-uniform.reject\endcsname{0.0}
\expandafter\def\csname ppt@rev.pol.boot.f20.fixedA.B4000.uiv.p1.rcr-vs-replay-ppi-uniform.reject\endcsname{0.0}
\expandafter\def\csname val@rev.pol.boot.f20.fixedA.B4000.uiv.p2.rcr-vs-generic-ppi-source.p-holm\endcsname{0.041}
\expandafter\def\csname cnt@rev.pol.boot.f20.fixedA.B4000.uiv.p2.rcr-vs-generic-ppi-source.p-holm\endcsname{0}
\expandafter\def\csname pct@rev.pol.boot.f20.fixedA.B4000.uiv.p2.rcr-vs-generic-ppi-source.p-holm\endcsname{4.1}
\expandafter\def\csname ppt@rev.pol.boot.f20.fixedA.B4000.uiv.p2.rcr-vs-generic-ppi-source.p-holm\endcsname{$+$4.1}
\expandafter\def\csname val@rev.pol.boot.f20.fixedA.B4000.uiv.p2.rcr-vs-generic-ppi-source.p-raw\endcsname{0.010}
\expandafter\def\csname cnt@rev.pol.boot.f20.fixedA.B4000.uiv.p2.rcr-vs-generic-ppi-source.p-raw\endcsname{0}
\expandafter\def\csname pct@rev.pol.boot.f20.fixedA.B4000.uiv.p2.rcr-vs-generic-ppi-source.p-raw\endcsname{1.0}
\expandafter\def\csname ppt@rev.pol.boot.f20.fixedA.B4000.uiv.p2.rcr-vs-generic-ppi-source.p-raw\endcsname{$+$1.0}
\expandafter\def\csname val@rev.pol.boot.f20.fixedA.B4000.uiv.p2.rcr-vs-generic-ppi-source.reject\endcsname{1}
\expandafter\def\csname cnt@rev.pol.boot.f20.fixedA.B4000.uiv.p2.rcr-vs-generic-ppi-source.reject\endcsname{1}
\expandafter\def\csname pct@rev.pol.boot.f20.fixedA.B4000.uiv.p2.rcr-vs-generic-ppi-source.reject\endcsname{100.0}
\expandafter\def\csname ppt@rev.pol.boot.f20.fixedA.B4000.uiv.p2.rcr-vs-generic-ppi-source.reject\endcsname{$+$100.0}
\expandafter\def\csname val@rev.pol.boot.f20.fixedA.B4000.uiv.p2.rcr-vs-replay-ppi-uniform.p-holm\endcsname{0.041}
\expandafter\def\csname cnt@rev.pol.boot.f20.fixedA.B4000.uiv.p2.rcr-vs-replay-ppi-uniform.p-holm\endcsname{0}
\expandafter\def\csname pct@rev.pol.boot.f20.fixedA.B4000.uiv.p2.rcr-vs-replay-ppi-uniform.p-holm\endcsname{4.1}
\expandafter\def\csname ppt@rev.pol.boot.f20.fixedA.B4000.uiv.p2.rcr-vs-replay-ppi-uniform.p-holm\endcsname{$+$4.1}
\expandafter\def\csname val@rev.pol.boot.f20.fixedA.B4000.uiv.p2.rcr-vs-replay-ppi-uniform.p-raw\endcsname{0.016}
\expandafter\def\csname cnt@rev.pol.boot.f20.fixedA.B4000.uiv.p2.rcr-vs-replay-ppi-uniform.p-raw\endcsname{0}
\expandafter\def\csname pct@rev.pol.boot.f20.fixedA.B4000.uiv.p2.rcr-vs-replay-ppi-uniform.p-raw\endcsname{1.6}
\expandafter\def\csname ppt@rev.pol.boot.f20.fixedA.B4000.uiv.p2.rcr-vs-replay-ppi-uniform.p-raw\endcsname{$+$1.6}
\expandafter\def\csname val@rev.pol.boot.f20.fixedA.B4000.uiv.p2.rcr-vs-replay-ppi-uniform.reject\endcsname{1}
\expandafter\def\csname cnt@rev.pol.boot.f20.fixedA.B4000.uiv.p2.rcr-vs-replay-ppi-uniform.reject\endcsname{1}
\expandafter\def\csname pct@rev.pol.boot.f20.fixedA.B4000.uiv.p2.rcr-vs-replay-ppi-uniform.reject\endcsname{100.0}
\expandafter\def\csname ppt@rev.pol.boot.f20.fixedA.B4000.uiv.p2.rcr-vs-replay-ppi-uniform.reject\endcsname{$+$100.0}
\expandafter\def\csname val@rev.pol.boot.f20.fixedB.B100000.max-p-holm-rejected\endcsname{0.046}
\expandafter\def\csname cnt@rev.pol.boot.f20.fixedB.B100000.max-p-holm-rejected\endcsname{0}
\expandafter\def\csname pct@rev.pol.boot.f20.fixedB.B100000.max-p-holm-rejected\endcsname{4.6}
\expandafter\def\csname ppt@rev.pol.boot.f20.fixedB.B100000.max-p-holm-rejected\endcsname{$+$4.6}
\expandafter\def\csname val@rev.pol.boot.f20.fixedB.B100000.n-rejected\endcsname{7}
\expandafter\def\csname cnt@rev.pol.boot.f20.fixedB.B100000.n-rejected\endcsname{7}
\expandafter\def\csname pct@rev.pol.boot.f20.fixedB.B100000.n-rejected\endcsname{700.0}
\expandafter\def\csname ppt@rev.pol.boot.f20.fixedB.B100000.n-rejected\endcsname{$+$700.0}
\expandafter\def\csname val@rev.pol.boot.f20.fixedB.B100000.sspro.p1.rcr-vs-generic-ppi-source.p-holm\endcsname{0.028}
\expandafter\def\csname cnt@rev.pol.boot.f20.fixedB.B100000.sspro.p1.rcr-vs-generic-ppi-source.p-holm\endcsname{0}
\expandafter\def\csname pct@rev.pol.boot.f20.fixedB.B100000.sspro.p1.rcr-vs-generic-ppi-source.p-holm\endcsname{2.8}
\expandafter\def\csname ppt@rev.pol.boot.f20.fixedB.B100000.sspro.p1.rcr-vs-generic-ppi-source.p-holm\endcsname{$+$2.8}
\expandafter\def\csname val@rev.pol.boot.f20.fixedB.B100000.sspro.p1.rcr-vs-generic-ppi-source.p-raw\endcsname{0.006}
\expandafter\def\csname cnt@rev.pol.boot.f20.fixedB.B100000.sspro.p1.rcr-vs-generic-ppi-source.p-raw\endcsname{0}
\expandafter\def\csname pct@rev.pol.boot.f20.fixedB.B100000.sspro.p1.rcr-vs-generic-ppi-source.p-raw\endcsname{0.6}
\expandafter\def\csname ppt@rev.pol.boot.f20.fixedB.B100000.sspro.p1.rcr-vs-generic-ppi-source.p-raw\endcsname{$+$0.6}
\expandafter\def\csname val@rev.pol.boot.f20.fixedB.B100000.sspro.p1.rcr-vs-generic-ppi-source.reject\endcsname{1}
\expandafter\def\csname cnt@rev.pol.boot.f20.fixedB.B100000.sspro.p1.rcr-vs-generic-ppi-source.reject\endcsname{1}
\expandafter\def\csname pct@rev.pol.boot.f20.fixedB.B100000.sspro.p1.rcr-vs-generic-ppi-source.reject\endcsname{100.0}
\expandafter\def\csname ppt@rev.pol.boot.f20.fixedB.B100000.sspro.p1.rcr-vs-generic-ppi-source.reject\endcsname{$+$100.0}
\expandafter\def\csname val@rev.pol.boot.f20.fixedB.B100000.sspro.p1.rcr-vs-replay-ppi-uniform.p-holm\endcsname{0.046}
\expandafter\def\csname cnt@rev.pol.boot.f20.fixedB.B100000.sspro.p1.rcr-vs-replay-ppi-uniform.p-holm\endcsname{0}
\expandafter\def\csname pct@rev.pol.boot.f20.fixedB.B100000.sspro.p1.rcr-vs-replay-ppi-uniform.p-holm\endcsname{4.6}
\expandafter\def\csname ppt@rev.pol.boot.f20.fixedB.B100000.sspro.p1.rcr-vs-replay-ppi-uniform.p-holm\endcsname{$+$4.6}
\expandafter\def\csname val@rev.pol.boot.f20.fixedB.B100000.sspro.p1.rcr-vs-replay-ppi-uniform.p-raw\endcsname{0.013}
\expandafter\def\csname cnt@rev.pol.boot.f20.fixedB.B100000.sspro.p1.rcr-vs-replay-ppi-uniform.p-raw\endcsname{0}
\expandafter\def\csname pct@rev.pol.boot.f20.fixedB.B100000.sspro.p1.rcr-vs-replay-ppi-uniform.p-raw\endcsname{1.3}
\expandafter\def\csname ppt@rev.pol.boot.f20.fixedB.B100000.sspro.p1.rcr-vs-replay-ppi-uniform.p-raw\endcsname{$+$1.3}
\expandafter\def\csname val@rev.pol.boot.f20.fixedB.B100000.sspro.p1.rcr-vs-replay-ppi-uniform.reject\endcsname{1}
\expandafter\def\csname cnt@rev.pol.boot.f20.fixedB.B100000.sspro.p1.rcr-vs-replay-ppi-uniform.reject\endcsname{1}
\expandafter\def\csname pct@rev.pol.boot.f20.fixedB.B100000.sspro.p1.rcr-vs-replay-ppi-uniform.reject\endcsname{100.0}
\expandafter\def\csname ppt@rev.pol.boot.f20.fixedB.B100000.sspro.p1.rcr-vs-replay-ppi-uniform.reject\endcsname{$+$100.0}
\expandafter\def\csname val@rev.pol.boot.f20.fixedB.B100000.sspro.p2.rcr-vs-generic-ppi-source.p-holm\endcsname{0.000}
\expandafter\def\csname cnt@rev.pol.boot.f20.fixedB.B100000.sspro.p2.rcr-vs-generic-ppi-source.p-holm\endcsname{0}
\expandafter\def\csname pct@rev.pol.boot.f20.fixedB.B100000.sspro.p2.rcr-vs-generic-ppi-source.p-holm\endcsname{0.0}
\expandafter\def\csname ppt@rev.pol.boot.f20.fixedB.B100000.sspro.p2.rcr-vs-generic-ppi-source.p-holm\endcsname{0.0}
\expandafter\def\csname val@rev.pol.boot.f20.fixedB.B100000.sspro.p2.rcr-vs-generic-ppi-source.p-raw\endcsname{0.000}
\expandafter\def\csname cnt@rev.pol.boot.f20.fixedB.B100000.sspro.p2.rcr-vs-generic-ppi-source.p-raw\endcsname{0}
\expandafter\def\csname pct@rev.pol.boot.f20.fixedB.B100000.sspro.p2.rcr-vs-generic-ppi-source.p-raw\endcsname{0.0}
\expandafter\def\csname ppt@rev.pol.boot.f20.fixedB.B100000.sspro.p2.rcr-vs-generic-ppi-source.p-raw\endcsname{0.0}
\expandafter\def\csname val@rev.pol.boot.f20.fixedB.B100000.sspro.p2.rcr-vs-generic-ppi-source.reject\endcsname{1}
\expandafter\def\csname cnt@rev.pol.boot.f20.fixedB.B100000.sspro.p2.rcr-vs-generic-ppi-source.reject\endcsname{1}
\expandafter\def\csname pct@rev.pol.boot.f20.fixedB.B100000.sspro.p2.rcr-vs-generic-ppi-source.reject\endcsname{100.0}
\expandafter\def\csname ppt@rev.pol.boot.f20.fixedB.B100000.sspro.p2.rcr-vs-generic-ppi-source.reject\endcsname{$+$100.0}
\expandafter\def\csname val@rev.pol.boot.f20.fixedB.B100000.sspro.p2.rcr-vs-replay-ppi-uniform.p-holm\endcsname{0.000}
\expandafter\def\csname cnt@rev.pol.boot.f20.fixedB.B100000.sspro.p2.rcr-vs-replay-ppi-uniform.p-holm\endcsname{0}
\expandafter\def\csname pct@rev.pol.boot.f20.fixedB.B100000.sspro.p2.rcr-vs-replay-ppi-uniform.p-holm\endcsname{0.0}
\expandafter\def\csname ppt@rev.pol.boot.f20.fixedB.B100000.sspro.p2.rcr-vs-replay-ppi-uniform.p-holm\endcsname{0.0}
\expandafter\def\csname val@rev.pol.boot.f20.fixedB.B100000.sspro.p2.rcr-vs-replay-ppi-uniform.p-raw\endcsname{0.000}
\expandafter\def\csname cnt@rev.pol.boot.f20.fixedB.B100000.sspro.p2.rcr-vs-replay-ppi-uniform.p-raw\endcsname{0}
\expandafter\def\csname pct@rev.pol.boot.f20.fixedB.B100000.sspro.p2.rcr-vs-replay-ppi-uniform.p-raw\endcsname{0.0}
\expandafter\def\csname ppt@rev.pol.boot.f20.fixedB.B100000.sspro.p2.rcr-vs-replay-ppi-uniform.p-raw\endcsname{0.0}
\expandafter\def\csname val@rev.pol.boot.f20.fixedB.B100000.sspro.p2.rcr-vs-replay-ppi-uniform.reject\endcsname{1}
\expandafter\def\csname cnt@rev.pol.boot.f20.fixedB.B100000.sspro.p2.rcr-vs-replay-ppi-uniform.reject\endcsname{1}
\expandafter\def\csname pct@rev.pol.boot.f20.fixedB.B100000.sspro.p2.rcr-vs-replay-ppi-uniform.reject\endcsname{100.0}
\expandafter\def\csname ppt@rev.pol.boot.f20.fixedB.B100000.sspro.p2.rcr-vs-replay-ppi-uniform.reject\endcsname{$+$100.0}
\expandafter\def\csname val@rev.pol.boot.f20.fixedB.B100000.uiv.p1.rcr-vs-generic-ppi-source.p-holm\endcsname{0.000}
\expandafter\def\csname cnt@rev.pol.boot.f20.fixedB.B100000.uiv.p1.rcr-vs-generic-ppi-source.p-holm\endcsname{0}
\expandafter\def\csname pct@rev.pol.boot.f20.fixedB.B100000.uiv.p1.rcr-vs-generic-ppi-source.p-holm\endcsname{0.0}
\expandafter\def\csname ppt@rev.pol.boot.f20.fixedB.B100000.uiv.p1.rcr-vs-generic-ppi-source.p-holm\endcsname{0.0}
\expandafter\def\csname val@rev.pol.boot.f20.fixedB.B100000.uiv.p1.rcr-vs-generic-ppi-source.p-raw\endcsname{0.000}
\expandafter\def\csname cnt@rev.pol.boot.f20.fixedB.B100000.uiv.p1.rcr-vs-generic-ppi-source.p-raw\endcsname{0}
\expandafter\def\csname pct@rev.pol.boot.f20.fixedB.B100000.uiv.p1.rcr-vs-generic-ppi-source.p-raw\endcsname{0.0}
\expandafter\def\csname ppt@rev.pol.boot.f20.fixedB.B100000.uiv.p1.rcr-vs-generic-ppi-source.p-raw\endcsname{0.0}
\expandafter\def\csname val@rev.pol.boot.f20.fixedB.B100000.uiv.p1.rcr-vs-generic-ppi-source.reject\endcsname{1}
\expandafter\def\csname cnt@rev.pol.boot.f20.fixedB.B100000.uiv.p1.rcr-vs-generic-ppi-source.reject\endcsname{1}
\expandafter\def\csname pct@rev.pol.boot.f20.fixedB.B100000.uiv.p1.rcr-vs-generic-ppi-source.reject\endcsname{100.0}
\expandafter\def\csname ppt@rev.pol.boot.f20.fixedB.B100000.uiv.p1.rcr-vs-generic-ppi-source.reject\endcsname{$+$100.0}
\expandafter\def\csname val@rev.pol.boot.f20.fixedB.B100000.uiv.p1.rcr-vs-replay-ppi-uniform.p-holm\endcsname{0.164}
\expandafter\def\csname cnt@rev.pol.boot.f20.fixedB.B100000.uiv.p1.rcr-vs-replay-ppi-uniform.p-holm\endcsname{0}
\expandafter\def\csname pct@rev.pol.boot.f20.fixedB.B100000.uiv.p1.rcr-vs-replay-ppi-uniform.p-holm\endcsname{16.4}
\expandafter\def\csname ppt@rev.pol.boot.f20.fixedB.B100000.uiv.p1.rcr-vs-replay-ppi-uniform.p-holm\endcsname{$+$16.4}
\expandafter\def\csname val@rev.pol.boot.f20.fixedB.B100000.uiv.p1.rcr-vs-replay-ppi-uniform.p-raw\endcsname{0.164}
\expandafter\def\csname cnt@rev.pol.boot.f20.fixedB.B100000.uiv.p1.rcr-vs-replay-ppi-uniform.p-raw\endcsname{0}
\expandafter\def\csname pct@rev.pol.boot.f20.fixedB.B100000.uiv.p1.rcr-vs-replay-ppi-uniform.p-raw\endcsname{16.4}
\expandafter\def\csname ppt@rev.pol.boot.f20.fixedB.B100000.uiv.p1.rcr-vs-replay-ppi-uniform.p-raw\endcsname{$+$16.4}
\expandafter\def\csname val@rev.pol.boot.f20.fixedB.B100000.uiv.p1.rcr-vs-replay-ppi-uniform.reject\endcsname{0}
\expandafter\def\csname cnt@rev.pol.boot.f20.fixedB.B100000.uiv.p1.rcr-vs-replay-ppi-uniform.reject\endcsname{0}
\expandafter\def\csname pct@rev.pol.boot.f20.fixedB.B100000.uiv.p1.rcr-vs-replay-ppi-uniform.reject\endcsname{0.0}
\expandafter\def\csname ppt@rev.pol.boot.f20.fixedB.B100000.uiv.p1.rcr-vs-replay-ppi-uniform.reject\endcsname{0.0}
\expandafter\def\csname val@rev.pol.boot.f20.fixedB.B100000.uiv.p2.rcr-vs-generic-ppi-source.p-holm\endcsname{0.046}
\expandafter\def\csname cnt@rev.pol.boot.f20.fixedB.B100000.uiv.p2.rcr-vs-generic-ppi-source.p-holm\endcsname{0}
\expandafter\def\csname pct@rev.pol.boot.f20.fixedB.B100000.uiv.p2.rcr-vs-generic-ppi-source.p-holm\endcsname{4.6}
\expandafter\def\csname ppt@rev.pol.boot.f20.fixedB.B100000.uiv.p2.rcr-vs-generic-ppi-source.p-holm\endcsname{$+$4.6}
\expandafter\def\csname val@rev.pol.boot.f20.fixedB.B100000.uiv.p2.rcr-vs-generic-ppi-source.p-raw\endcsname{0.011}
\expandafter\def\csname cnt@rev.pol.boot.f20.fixedB.B100000.uiv.p2.rcr-vs-generic-ppi-source.p-raw\endcsname{0}
\expandafter\def\csname pct@rev.pol.boot.f20.fixedB.B100000.uiv.p2.rcr-vs-generic-ppi-source.p-raw\endcsname{1.1}
\expandafter\def\csname ppt@rev.pol.boot.f20.fixedB.B100000.uiv.p2.rcr-vs-generic-ppi-source.p-raw\endcsname{$+$1.1}
\expandafter\def\csname val@rev.pol.boot.f20.fixedB.B100000.uiv.p2.rcr-vs-generic-ppi-source.reject\endcsname{1}
\expandafter\def\csname cnt@rev.pol.boot.f20.fixedB.B100000.uiv.p2.rcr-vs-generic-ppi-source.reject\endcsname{1}
\expandafter\def\csname pct@rev.pol.boot.f20.fixedB.B100000.uiv.p2.rcr-vs-generic-ppi-source.reject\endcsname{100.0}
\expandafter\def\csname ppt@rev.pol.boot.f20.fixedB.B100000.uiv.p2.rcr-vs-generic-ppi-source.reject\endcsname{$+$100.0}
\expandafter\def\csname val@rev.pol.boot.f20.fixedB.B100000.uiv.p2.rcr-vs-replay-ppi-uniform.p-holm\endcsname{0.046}
\expandafter\def\csname cnt@rev.pol.boot.f20.fixedB.B100000.uiv.p2.rcr-vs-replay-ppi-uniform.p-holm\endcsname{0}
\expandafter\def\csname pct@rev.pol.boot.f20.fixedB.B100000.uiv.p2.rcr-vs-replay-ppi-uniform.p-holm\endcsname{4.6}
\expandafter\def\csname ppt@rev.pol.boot.f20.fixedB.B100000.uiv.p2.rcr-vs-replay-ppi-uniform.p-holm\endcsname{$+$4.6}
\expandafter\def\csname val@rev.pol.boot.f20.fixedB.B100000.uiv.p2.rcr-vs-replay-ppi-uniform.p-raw\endcsname{0.013}
\expandafter\def\csname cnt@rev.pol.boot.f20.fixedB.B100000.uiv.p2.rcr-vs-replay-ppi-uniform.p-raw\endcsname{0}
\expandafter\def\csname pct@rev.pol.boot.f20.fixedB.B100000.uiv.p2.rcr-vs-replay-ppi-uniform.p-raw\endcsname{1.3}
\expandafter\def\csname ppt@rev.pol.boot.f20.fixedB.B100000.uiv.p2.rcr-vs-replay-ppi-uniform.p-raw\endcsname{$+$1.3}
\expandafter\def\csname val@rev.pol.boot.f20.fixedB.B100000.uiv.p2.rcr-vs-replay-ppi-uniform.reject\endcsname{1}
\expandafter\def\csname cnt@rev.pol.boot.f20.fixedB.B100000.uiv.p2.rcr-vs-replay-ppi-uniform.reject\endcsname{1}
\expandafter\def\csname pct@rev.pol.boot.f20.fixedB.B100000.uiv.p2.rcr-vs-replay-ppi-uniform.reject\endcsname{100.0}
\expandafter\def\csname ppt@rev.pol.boot.f20.fixedB.B100000.uiv.p2.rcr-vs-replay-ppi-uniform.reject\endcsname{$+$100.0}
\expandafter\def\csname val@rev.pol.boot.f20.fixedB.B4000.n-rejected\endcsname{7}
\expandafter\def\csname cnt@rev.pol.boot.f20.fixedB.B4000.n-rejected\endcsname{7}
\expandafter\def\csname pct@rev.pol.boot.f20.fixedB.B4000.n-rejected\endcsname{700.0}
\expandafter\def\csname ppt@rev.pol.boot.f20.fixedB.B4000.n-rejected\endcsname{$+$700.0}
\expandafter\def\csname val@rev.pol.boot.f20.fixedB.B4000.sspro.p1.rcr-vs-generic-ppi-source.p-holm\endcsname{0.036}
\expandafter\def\csname cnt@rev.pol.boot.f20.fixedB.B4000.sspro.p1.rcr-vs-generic-ppi-source.p-holm\endcsname{0}
\expandafter\def\csname pct@rev.pol.boot.f20.fixedB.B4000.sspro.p1.rcr-vs-generic-ppi-source.p-holm\endcsname{3.6}
\expandafter\def\csname ppt@rev.pol.boot.f20.fixedB.B4000.sspro.p1.rcr-vs-generic-ppi-source.p-holm\endcsname{$+$3.6}
\expandafter\def\csname val@rev.pol.boot.f20.fixedB.B4000.sspro.p1.rcr-vs-generic-ppi-source.p-raw\endcsname{0.007}
\expandafter\def\csname cnt@rev.pol.boot.f20.fixedB.B4000.sspro.p1.rcr-vs-generic-ppi-source.p-raw\endcsname{0}
\expandafter\def\csname pct@rev.pol.boot.f20.fixedB.B4000.sspro.p1.rcr-vs-generic-ppi-source.p-raw\endcsname{0.7}
\expandafter\def\csname ppt@rev.pol.boot.f20.fixedB.B4000.sspro.p1.rcr-vs-generic-ppi-source.p-raw\endcsname{$+$0.7}
\expandafter\def\csname val@rev.pol.boot.f20.fixedB.B4000.sspro.p1.rcr-vs-generic-ppi-source.reject\endcsname{1}
\expandafter\def\csname cnt@rev.pol.boot.f20.fixedB.B4000.sspro.p1.rcr-vs-generic-ppi-source.reject\endcsname{1}
\expandafter\def\csname pct@rev.pol.boot.f20.fixedB.B4000.sspro.p1.rcr-vs-generic-ppi-source.reject\endcsname{100.0}
\expandafter\def\csname ppt@rev.pol.boot.f20.fixedB.B4000.sspro.p1.rcr-vs-generic-ppi-source.reject\endcsname{$+$100.0}
\expandafter\def\csname val@rev.pol.boot.f20.fixedB.B4000.sspro.p1.rcr-vs-replay-ppi-uniform.p-holm\endcsname{0.043}
\expandafter\def\csname cnt@rev.pol.boot.f20.fixedB.B4000.sspro.p1.rcr-vs-replay-ppi-uniform.p-holm\endcsname{0}
\expandafter\def\csname pct@rev.pol.boot.f20.fixedB.B4000.sspro.p1.rcr-vs-replay-ppi-uniform.p-holm\endcsname{4.3}
\expandafter\def\csname ppt@rev.pol.boot.f20.fixedB.B4000.sspro.p1.rcr-vs-replay-ppi-uniform.p-holm\endcsname{$+$4.3}
\expandafter\def\csname val@rev.pol.boot.f20.fixedB.B4000.sspro.p1.rcr-vs-replay-ppi-uniform.p-raw\endcsname{0.014}
\expandafter\def\csname cnt@rev.pol.boot.f20.fixedB.B4000.sspro.p1.rcr-vs-replay-ppi-uniform.p-raw\endcsname{0}
\expandafter\def\csname pct@rev.pol.boot.f20.fixedB.B4000.sspro.p1.rcr-vs-replay-ppi-uniform.p-raw\endcsname{1.4}
\expandafter\def\csname ppt@rev.pol.boot.f20.fixedB.B4000.sspro.p1.rcr-vs-replay-ppi-uniform.p-raw\endcsname{$+$1.4}
\expandafter\def\csname val@rev.pol.boot.f20.fixedB.B4000.sspro.p1.rcr-vs-replay-ppi-uniform.reject\endcsname{1}
\expandafter\def\csname cnt@rev.pol.boot.f20.fixedB.B4000.sspro.p1.rcr-vs-replay-ppi-uniform.reject\endcsname{1}
\expandafter\def\csname pct@rev.pol.boot.f20.fixedB.B4000.sspro.p1.rcr-vs-replay-ppi-uniform.reject\endcsname{100.0}
\expandafter\def\csname ppt@rev.pol.boot.f20.fixedB.B4000.sspro.p1.rcr-vs-replay-ppi-uniform.reject\endcsname{$+$100.0}
\expandafter\def\csname val@rev.pol.boot.f20.fixedB.B4000.sspro.p2.rcr-vs-generic-ppi-source.p-holm\endcsname{0.002}
\expandafter\def\csname cnt@rev.pol.boot.f20.fixedB.B4000.sspro.p2.rcr-vs-generic-ppi-source.p-holm\endcsname{0}
\expandafter\def\csname pct@rev.pol.boot.f20.fixedB.B4000.sspro.p2.rcr-vs-generic-ppi-source.p-holm\endcsname{0.2}
\expandafter\def\csname ppt@rev.pol.boot.f20.fixedB.B4000.sspro.p2.rcr-vs-generic-ppi-source.p-holm\endcsname{$+$0.2}
\expandafter\def\csname val@rev.pol.boot.f20.fixedB.B4000.sspro.p2.rcr-vs-generic-ppi-source.p-raw\endcsname{0.000}
\expandafter\def\csname cnt@rev.pol.boot.f20.fixedB.B4000.sspro.p2.rcr-vs-generic-ppi-source.p-raw\endcsname{0}
\expandafter\def\csname pct@rev.pol.boot.f20.fixedB.B4000.sspro.p2.rcr-vs-generic-ppi-source.p-raw\endcsname{0.0}
\expandafter\def\csname ppt@rev.pol.boot.f20.fixedB.B4000.sspro.p2.rcr-vs-generic-ppi-source.p-raw\endcsname{0.0}
\expandafter\def\csname val@rev.pol.boot.f20.fixedB.B4000.sspro.p2.rcr-vs-generic-ppi-source.reject\endcsname{1}
\expandafter\def\csname cnt@rev.pol.boot.f20.fixedB.B4000.sspro.p2.rcr-vs-generic-ppi-source.reject\endcsname{1}
\expandafter\def\csname pct@rev.pol.boot.f20.fixedB.B4000.sspro.p2.rcr-vs-generic-ppi-source.reject\endcsname{100.0}
\expandafter\def\csname ppt@rev.pol.boot.f20.fixedB.B4000.sspro.p2.rcr-vs-generic-ppi-source.reject\endcsname{$+$100.0}
\expandafter\def\csname val@rev.pol.boot.f20.fixedB.B4000.sspro.p2.rcr-vs-replay-ppi-uniform.p-holm\endcsname{0.002}
\expandafter\def\csname cnt@rev.pol.boot.f20.fixedB.B4000.sspro.p2.rcr-vs-replay-ppi-uniform.p-holm\endcsname{0}
\expandafter\def\csname pct@rev.pol.boot.f20.fixedB.B4000.sspro.p2.rcr-vs-replay-ppi-uniform.p-holm\endcsname{0.2}
\expandafter\def\csname ppt@rev.pol.boot.f20.fixedB.B4000.sspro.p2.rcr-vs-replay-ppi-uniform.p-holm\endcsname{$+$0.2}
\expandafter\def\csname val@rev.pol.boot.f20.fixedB.B4000.sspro.p2.rcr-vs-replay-ppi-uniform.p-raw\endcsname{0.000}
\expandafter\def\csname cnt@rev.pol.boot.f20.fixedB.B4000.sspro.p2.rcr-vs-replay-ppi-uniform.p-raw\endcsname{0}
\expandafter\def\csname pct@rev.pol.boot.f20.fixedB.B4000.sspro.p2.rcr-vs-replay-ppi-uniform.p-raw\endcsname{0.0}
\expandafter\def\csname ppt@rev.pol.boot.f20.fixedB.B4000.sspro.p2.rcr-vs-replay-ppi-uniform.p-raw\endcsname{0.0}
\expandafter\def\csname val@rev.pol.boot.f20.fixedB.B4000.sspro.p2.rcr-vs-replay-ppi-uniform.reject\endcsname{1}
\expandafter\def\csname cnt@rev.pol.boot.f20.fixedB.B4000.sspro.p2.rcr-vs-replay-ppi-uniform.reject\endcsname{1}
\expandafter\def\csname pct@rev.pol.boot.f20.fixedB.B4000.sspro.p2.rcr-vs-replay-ppi-uniform.reject\endcsname{100.0}
\expandafter\def\csname ppt@rev.pol.boot.f20.fixedB.B4000.sspro.p2.rcr-vs-replay-ppi-uniform.reject\endcsname{$+$100.0}
\expandafter\def\csname val@rev.pol.boot.f20.fixedB.B4000.uiv.p1.rcr-vs-generic-ppi-source.p-holm\endcsname{0.002}
\expandafter\def\csname cnt@rev.pol.boot.f20.fixedB.B4000.uiv.p1.rcr-vs-generic-ppi-source.p-holm\endcsname{0}
\expandafter\def\csname pct@rev.pol.boot.f20.fixedB.B4000.uiv.p1.rcr-vs-generic-ppi-source.p-holm\endcsname{0.2}
\expandafter\def\csname ppt@rev.pol.boot.f20.fixedB.B4000.uiv.p1.rcr-vs-generic-ppi-source.p-holm\endcsname{$+$0.2}
\expandafter\def\csname val@rev.pol.boot.f20.fixedB.B4000.uiv.p1.rcr-vs-generic-ppi-source.p-raw\endcsname{0.000}
\expandafter\def\csname cnt@rev.pol.boot.f20.fixedB.B4000.uiv.p1.rcr-vs-generic-ppi-source.p-raw\endcsname{0}
\expandafter\def\csname pct@rev.pol.boot.f20.fixedB.B4000.uiv.p1.rcr-vs-generic-ppi-source.p-raw\endcsname{0.0}
\expandafter\def\csname ppt@rev.pol.boot.f20.fixedB.B4000.uiv.p1.rcr-vs-generic-ppi-source.p-raw\endcsname{0.0}
\expandafter\def\csname val@rev.pol.boot.f20.fixedB.B4000.uiv.p1.rcr-vs-generic-ppi-source.reject\endcsname{1}
\expandafter\def\csname cnt@rev.pol.boot.f20.fixedB.B4000.uiv.p1.rcr-vs-generic-ppi-source.reject\endcsname{1}
\expandafter\def\csname pct@rev.pol.boot.f20.fixedB.B4000.uiv.p1.rcr-vs-generic-ppi-source.reject\endcsname{100.0}
\expandafter\def\csname ppt@rev.pol.boot.f20.fixedB.B4000.uiv.p1.rcr-vs-generic-ppi-source.reject\endcsname{$+$100.0}
\expandafter\def\csname val@rev.pol.boot.f20.fixedB.B4000.uiv.p1.rcr-vs-replay-ppi-uniform.p-holm\endcsname{0.166}
\expandafter\def\csname cnt@rev.pol.boot.f20.fixedB.B4000.uiv.p1.rcr-vs-replay-ppi-uniform.p-holm\endcsname{0}
\expandafter\def\csname pct@rev.pol.boot.f20.fixedB.B4000.uiv.p1.rcr-vs-replay-ppi-uniform.p-holm\endcsname{16.6}
\expandafter\def\csname ppt@rev.pol.boot.f20.fixedB.B4000.uiv.p1.rcr-vs-replay-ppi-uniform.p-holm\endcsname{$+$16.6}
\expandafter\def\csname val@rev.pol.boot.f20.fixedB.B4000.uiv.p1.rcr-vs-replay-ppi-uniform.p-raw\endcsname{0.166}
\expandafter\def\csname cnt@rev.pol.boot.f20.fixedB.B4000.uiv.p1.rcr-vs-replay-ppi-uniform.p-raw\endcsname{0}
\expandafter\def\csname pct@rev.pol.boot.f20.fixedB.B4000.uiv.p1.rcr-vs-replay-ppi-uniform.p-raw\endcsname{16.6}
\expandafter\def\csname ppt@rev.pol.boot.f20.fixedB.B4000.uiv.p1.rcr-vs-replay-ppi-uniform.p-raw\endcsname{$+$16.6}
\expandafter\def\csname val@rev.pol.boot.f20.fixedB.B4000.uiv.p1.rcr-vs-replay-ppi-uniform.reject\endcsname{0}
\expandafter\def\csname cnt@rev.pol.boot.f20.fixedB.B4000.uiv.p1.rcr-vs-replay-ppi-uniform.reject\endcsname{0}
\expandafter\def\csname pct@rev.pol.boot.f20.fixedB.B4000.uiv.p1.rcr-vs-replay-ppi-uniform.reject\endcsname{0.0}
\expandafter\def\csname ppt@rev.pol.boot.f20.fixedB.B4000.uiv.p1.rcr-vs-replay-ppi-uniform.reject\endcsname{0.0}
\expandafter\def\csname val@rev.pol.boot.f20.fixedB.B4000.uiv.p2.rcr-vs-generic-ppi-source.p-holm\endcsname{0.042}
\expandafter\def\csname cnt@rev.pol.boot.f20.fixedB.B4000.uiv.p2.rcr-vs-generic-ppi-source.p-holm\endcsname{0}
\expandafter\def\csname pct@rev.pol.boot.f20.fixedB.B4000.uiv.p2.rcr-vs-generic-ppi-source.p-holm\endcsname{4.2}
\expandafter\def\csname ppt@rev.pol.boot.f20.fixedB.B4000.uiv.p2.rcr-vs-generic-ppi-source.p-holm\endcsname{$+$4.2}
\expandafter\def\csname val@rev.pol.boot.f20.fixedB.B4000.uiv.p2.rcr-vs-generic-ppi-source.p-raw\endcsname{0.010}
\expandafter\def\csname cnt@rev.pol.boot.f20.fixedB.B4000.uiv.p2.rcr-vs-generic-ppi-source.p-raw\endcsname{0}
\expandafter\def\csname pct@rev.pol.boot.f20.fixedB.B4000.uiv.p2.rcr-vs-generic-ppi-source.p-raw\endcsname{1.0}
\expandafter\def\csname ppt@rev.pol.boot.f20.fixedB.B4000.uiv.p2.rcr-vs-generic-ppi-source.p-raw\endcsname{$+$1.0}
\expandafter\def\csname val@rev.pol.boot.f20.fixedB.B4000.uiv.p2.rcr-vs-generic-ppi-source.reject\endcsname{1}
\expandafter\def\csname cnt@rev.pol.boot.f20.fixedB.B4000.uiv.p2.rcr-vs-generic-ppi-source.reject\endcsname{1}
\expandafter\def\csname pct@rev.pol.boot.f20.fixedB.B4000.uiv.p2.rcr-vs-generic-ppi-source.reject\endcsname{100.0}
\expandafter\def\csname ppt@rev.pol.boot.f20.fixedB.B4000.uiv.p2.rcr-vs-generic-ppi-source.reject\endcsname{$+$100.0}
\expandafter\def\csname val@rev.pol.boot.f20.fixedB.B4000.uiv.p2.rcr-vs-replay-ppi-uniform.p-holm\endcsname{0.043}
\expandafter\def\csname cnt@rev.pol.boot.f20.fixedB.B4000.uiv.p2.rcr-vs-replay-ppi-uniform.p-holm\endcsname{0}
\expandafter\def\csname pct@rev.pol.boot.f20.fixedB.B4000.uiv.p2.rcr-vs-replay-ppi-uniform.p-holm\endcsname{4.3}
\expandafter\def\csname ppt@rev.pol.boot.f20.fixedB.B4000.uiv.p2.rcr-vs-replay-ppi-uniform.p-holm\endcsname{$+$4.3}
\expandafter\def\csname val@rev.pol.boot.f20.fixedB.B4000.uiv.p2.rcr-vs-replay-ppi-uniform.p-raw\endcsname{0.015}
\expandafter\def\csname cnt@rev.pol.boot.f20.fixedB.B4000.uiv.p2.rcr-vs-replay-ppi-uniform.p-raw\endcsname{0}
\expandafter\def\csname pct@rev.pol.boot.f20.fixedB.B4000.uiv.p2.rcr-vs-replay-ppi-uniform.p-raw\endcsname{1.5}
\expandafter\def\csname ppt@rev.pol.boot.f20.fixedB.B4000.uiv.p2.rcr-vs-replay-ppi-uniform.p-raw\endcsname{$+$1.5}
\expandafter\def\csname val@rev.pol.boot.f20.fixedB.B4000.uiv.p2.rcr-vs-replay-ppi-uniform.reject\endcsname{1}
\expandafter\def\csname cnt@rev.pol.boot.f20.fixedB.B4000.uiv.p2.rcr-vs-replay-ppi-uniform.reject\endcsname{1}
\expandafter\def\csname pct@rev.pol.boot.f20.fixedB.B4000.uiv.p2.rcr-vs-replay-ppi-uniform.reject\endcsname{100.0}
\expandafter\def\csname ppt@rev.pol.boot.f20.fixedB.B4000.uiv.p2.rcr-vs-replay-ppi-uniform.reject\endcsname{$+$100.0}
\expandafter\def\csname val@rev.pol.boot.n.B100000.seed-dependent\endcsname{0}
\expandafter\def\csname cnt@rev.pol.boot.n.B100000.seed-dependent\endcsname{0}
\expandafter\def\csname pct@rev.pol.boot.n.B100000.seed-dependent\endcsname{0.0}
\expandafter\def\csname ppt@rev.pol.boot.n.B100000.seed-dependent\endcsname{0.0}
\expandafter\def\csname val@rev.pol.boot.n.B100000.sspro.p1.rcr-vs-generic-ppi-source.p-holm-max\endcsname{0.028}
\expandafter\def\csname cnt@rev.pol.boot.n.B100000.sspro.p1.rcr-vs-generic-ppi-source.p-holm-max\endcsname{0}
\expandafter\def\csname pct@rev.pol.boot.n.B100000.sspro.p1.rcr-vs-generic-ppi-source.p-holm-max\endcsname{2.8}
\expandafter\def\csname ppt@rev.pol.boot.n.B100000.sspro.p1.rcr-vs-generic-ppi-source.p-holm-max\endcsname{$+$2.8}
\expandafter\def\csname val@rev.pol.boot.n.B100000.sspro.p1.rcr-vs-generic-ppi-source.p-holm-min\endcsname{0.027}
\expandafter\def\csname cnt@rev.pol.boot.n.B100000.sspro.p1.rcr-vs-generic-ppi-source.p-holm-min\endcsname{0}
\expandafter\def\csname pct@rev.pol.boot.n.B100000.sspro.p1.rcr-vs-generic-ppi-source.p-holm-min\endcsname{2.7}
\expandafter\def\csname ppt@rev.pol.boot.n.B100000.sspro.p1.rcr-vs-generic-ppi-source.p-holm-min\endcsname{$+$2.7}
\expandafter\def\csname val@rev.pol.boot.n.B100000.sspro.p1.rcr-vs-replay-ppi-uniform.p-holm-max\endcsname{0.037}
\expandafter\def\csname cnt@rev.pol.boot.n.B100000.sspro.p1.rcr-vs-replay-ppi-uniform.p-holm-max\endcsname{0}
\expandafter\def\csname pct@rev.pol.boot.n.B100000.sspro.p1.rcr-vs-replay-ppi-uniform.p-holm-max\endcsname{3.7}
\expandafter\def\csname ppt@rev.pol.boot.n.B100000.sspro.p1.rcr-vs-replay-ppi-uniform.p-holm-max\endcsname{$+$3.7}
\expandafter\def\csname val@rev.pol.boot.n.B100000.sspro.p1.rcr-vs-replay-ppi-uniform.p-holm-min\endcsname{0.036}
\expandafter\def\csname cnt@rev.pol.boot.n.B100000.sspro.p1.rcr-vs-replay-ppi-uniform.p-holm-min\endcsname{0}
\expandafter\def\csname pct@rev.pol.boot.n.B100000.sspro.p1.rcr-vs-replay-ppi-uniform.p-holm-min\endcsname{3.6}
\expandafter\def\csname ppt@rev.pol.boot.n.B100000.sspro.p1.rcr-vs-replay-ppi-uniform.p-holm-min\endcsname{$+$3.6}
\expandafter\def\csname val@rev.pol.boot.n.B100000.sspro.p2.rcr-vs-generic-ppi-source.p-holm-max\endcsname{0.893}
\expandafter\def\csname cnt@rev.pol.boot.n.B100000.sspro.p2.rcr-vs-generic-ppi-source.p-holm-max\endcsname{1}
\expandafter\def\csname pct@rev.pol.boot.n.B100000.sspro.p2.rcr-vs-generic-ppi-source.p-holm-max\endcsname{89.3}
\expandafter\def\csname ppt@rev.pol.boot.n.B100000.sspro.p2.rcr-vs-generic-ppi-source.p-holm-max\endcsname{$+$89.3}
\expandafter\def\csname val@rev.pol.boot.n.B100000.sspro.p2.rcr-vs-generic-ppi-source.p-holm-min\endcsname{0.889}
\expandafter\def\csname cnt@rev.pol.boot.n.B100000.sspro.p2.rcr-vs-generic-ppi-source.p-holm-min\endcsname{1}
\expandafter\def\csname pct@rev.pol.boot.n.B100000.sspro.p2.rcr-vs-generic-ppi-source.p-holm-min\endcsname{88.9}
\expandafter\def\csname ppt@rev.pol.boot.n.B100000.sspro.p2.rcr-vs-generic-ppi-source.p-holm-min\endcsname{$+$88.9}
\expandafter\def\csname val@rev.pol.boot.n.B100000.sspro.p2.rcr-vs-replay-ppi-uniform.p-holm-max\endcsname{0.893}
\expandafter\def\csname cnt@rev.pol.boot.n.B100000.sspro.p2.rcr-vs-replay-ppi-uniform.p-holm-max\endcsname{1}
\expandafter\def\csname pct@rev.pol.boot.n.B100000.sspro.p2.rcr-vs-replay-ppi-uniform.p-holm-max\endcsname{89.3}
\expandafter\def\csname ppt@rev.pol.boot.n.B100000.sspro.p2.rcr-vs-replay-ppi-uniform.p-holm-max\endcsname{$+$89.3}
\expandafter\def\csname val@rev.pol.boot.n.B100000.sspro.p2.rcr-vs-replay-ppi-uniform.p-holm-min\endcsname{0.889}
\expandafter\def\csname cnt@rev.pol.boot.n.B100000.sspro.p2.rcr-vs-replay-ppi-uniform.p-holm-min\endcsname{1}
\expandafter\def\csname pct@rev.pol.boot.n.B100000.sspro.p2.rcr-vs-replay-ppi-uniform.p-holm-min\endcsname{88.9}
\expandafter\def\csname ppt@rev.pol.boot.n.B100000.sspro.p2.rcr-vs-replay-ppi-uniform.p-holm-min\endcsname{$+$88.9}
\expandafter\def\csname val@rev.pol.boot.n.B100000.uiv.p1.rcr-vs-generic-ppi-source.p-holm-max\endcsname{0.001}
\expandafter\def\csname cnt@rev.pol.boot.n.B100000.uiv.p1.rcr-vs-generic-ppi-source.p-holm-max\endcsname{0}
\expandafter\def\csname pct@rev.pol.boot.n.B100000.uiv.p1.rcr-vs-generic-ppi-source.p-holm-max\endcsname{0.1}
\expandafter\def\csname ppt@rev.pol.boot.n.B100000.uiv.p1.rcr-vs-generic-ppi-source.p-holm-max\endcsname{$+$0.1}
\expandafter\def\csname val@rev.pol.boot.n.B100000.uiv.p1.rcr-vs-generic-ppi-source.p-holm-min\endcsname{0.000}
\expandafter\def\csname cnt@rev.pol.boot.n.B100000.uiv.p1.rcr-vs-generic-ppi-source.p-holm-min\endcsname{0}
\expandafter\def\csname pct@rev.pol.boot.n.B100000.uiv.p1.rcr-vs-generic-ppi-source.p-holm-min\endcsname{0.0}
\expandafter\def\csname ppt@rev.pol.boot.n.B100000.uiv.p1.rcr-vs-generic-ppi-source.p-holm-min\endcsname{0.0}
\expandafter\def\csname val@rev.pol.boot.n.B100000.uiv.p1.rcr-vs-replay-ppi-uniform.p-holm-max\endcsname{0.000}
\expandafter\def\csname cnt@rev.pol.boot.n.B100000.uiv.p1.rcr-vs-replay-ppi-uniform.p-holm-max\endcsname{0}
\expandafter\def\csname pct@rev.pol.boot.n.B100000.uiv.p1.rcr-vs-replay-ppi-uniform.p-holm-max\endcsname{0.0}
\expandafter\def\csname ppt@rev.pol.boot.n.B100000.uiv.p1.rcr-vs-replay-ppi-uniform.p-holm-max\endcsname{0.0}
\expandafter\def\csname val@rev.pol.boot.n.B100000.uiv.p1.rcr-vs-replay-ppi-uniform.p-holm-min\endcsname{0.000}
\expandafter\def\csname cnt@rev.pol.boot.n.B100000.uiv.p1.rcr-vs-replay-ppi-uniform.p-holm-min\endcsname{0}
\expandafter\def\csname pct@rev.pol.boot.n.B100000.uiv.p1.rcr-vs-replay-ppi-uniform.p-holm-min\endcsname{0.0}
\expandafter\def\csname ppt@rev.pol.boot.n.B100000.uiv.p1.rcr-vs-replay-ppi-uniform.p-holm-min\endcsname{0.0}
\expandafter\def\csname val@rev.pol.boot.n.B100000.uiv.p2.rcr-vs-generic-ppi-source.p-holm-max\endcsname{0.134}
\expandafter\def\csname cnt@rev.pol.boot.n.B100000.uiv.p2.rcr-vs-generic-ppi-source.p-holm-max\endcsname{0}
\expandafter\def\csname pct@rev.pol.boot.n.B100000.uiv.p2.rcr-vs-generic-ppi-source.p-holm-max\endcsname{13.4}
\expandafter\def\csname ppt@rev.pol.boot.n.B100000.uiv.p2.rcr-vs-generic-ppi-source.p-holm-max\endcsname{$+$13.4}
\expandafter\def\csname val@rev.pol.boot.n.B100000.uiv.p2.rcr-vs-generic-ppi-source.p-holm-min\endcsname{0.130}
\expandafter\def\csname cnt@rev.pol.boot.n.B100000.uiv.p2.rcr-vs-generic-ppi-source.p-holm-min\endcsname{0}
\expandafter\def\csname pct@rev.pol.boot.n.B100000.uiv.p2.rcr-vs-generic-ppi-source.p-holm-min\endcsname{13.0}
\expandafter\def\csname ppt@rev.pol.boot.n.B100000.uiv.p2.rcr-vs-generic-ppi-source.p-holm-min\endcsname{$+$13.0}
\expandafter\def\csname val@rev.pol.boot.n.B100000.uiv.p2.rcr-vs-replay-ppi-uniform.p-holm-max\endcsname{0.020}
\expandafter\def\csname cnt@rev.pol.boot.n.B100000.uiv.p2.rcr-vs-replay-ppi-uniform.p-holm-max\endcsname{0}
\expandafter\def\csname pct@rev.pol.boot.n.B100000.uiv.p2.rcr-vs-replay-ppi-uniform.p-holm-max\endcsname{2.0}
\expandafter\def\csname ppt@rev.pol.boot.n.B100000.uiv.p2.rcr-vs-replay-ppi-uniform.p-holm-max\endcsname{$+$2.0}
\expandafter\def\csname val@rev.pol.boot.n.B100000.uiv.p2.rcr-vs-replay-ppi-uniform.p-holm-min\endcsname{0.017}
\expandafter\def\csname cnt@rev.pol.boot.n.B100000.uiv.p2.rcr-vs-replay-ppi-uniform.p-holm-min\endcsname{0}
\expandafter\def\csname pct@rev.pol.boot.n.B100000.uiv.p2.rcr-vs-replay-ppi-uniform.p-holm-min\endcsname{1.7}
\expandafter\def\csname ppt@rev.pol.boot.n.B100000.uiv.p2.rcr-vs-replay-ppi-uniform.p-holm-min\endcsname{$+$1.7}
\expandafter\def\csname val@rev.pol.boot.n.B4000.seed-dependent\endcsname{0}
\expandafter\def\csname cnt@rev.pol.boot.n.B4000.seed-dependent\endcsname{0}
\expandafter\def\csname pct@rev.pol.boot.n.B4000.seed-dependent\endcsname{0.0}
\expandafter\def\csname ppt@rev.pol.boot.n.B4000.seed-dependent\endcsname{0.0}
\expandafter\def\csname val@rev.pol.boot.n.content.B100000.max-p-holm-rejected\endcsname{0.036}
\expandafter\def\csname cnt@rev.pol.boot.n.content.B100000.max-p-holm-rejected\endcsname{0}
\expandafter\def\csname pct@rev.pol.boot.n.content.B100000.max-p-holm-rejected\endcsname{3.6}
\expandafter\def\csname ppt@rev.pol.boot.n.content.B100000.max-p-holm-rejected\endcsname{$+$3.6}
\expandafter\def\csname val@rev.pol.boot.n.content.B100000.n-rejected\endcsname{5}
\expandafter\def\csname cnt@rev.pol.boot.n.content.B100000.n-rejected\endcsname{5}
\expandafter\def\csname pct@rev.pol.boot.n.content.B100000.n-rejected\endcsname{500.0}
\expandafter\def\csname ppt@rev.pol.boot.n.content.B100000.n-rejected\endcsname{$+$500.0}
\expandafter\def\csname val@rev.pol.boot.n.content.B100000.sspro.p1.rcr-vs-generic-ppi-source.p-holm\endcsname{0.027}
\expandafter\def\csname cnt@rev.pol.boot.n.content.B100000.sspro.p1.rcr-vs-generic-ppi-source.p-holm\endcsname{0}
\expandafter\def\csname pct@rev.pol.boot.n.content.B100000.sspro.p1.rcr-vs-generic-ppi-source.p-holm\endcsname{2.7}
\expandafter\def\csname ppt@rev.pol.boot.n.content.B100000.sspro.p1.rcr-vs-generic-ppi-source.p-holm\endcsname{$+$2.7}
\expandafter\def\csname val@rev.pol.boot.n.content.B100000.sspro.p1.rcr-vs-generic-ppi-source.p-raw\endcsname{0.005}
\expandafter\def\csname cnt@rev.pol.boot.n.content.B100000.sspro.p1.rcr-vs-generic-ppi-source.p-raw\endcsname{0}
\expandafter\def\csname pct@rev.pol.boot.n.content.B100000.sspro.p1.rcr-vs-generic-ppi-source.p-raw\endcsname{0.5}
\expandafter\def\csname ppt@rev.pol.boot.n.content.B100000.sspro.p1.rcr-vs-generic-ppi-source.p-raw\endcsname{$+$0.5}
\expandafter\def\csname val@rev.pol.boot.n.content.B100000.sspro.p1.rcr-vs-generic-ppi-source.reject\endcsname{1}
\expandafter\def\csname cnt@rev.pol.boot.n.content.B100000.sspro.p1.rcr-vs-generic-ppi-source.reject\endcsname{1}
\expandafter\def\csname pct@rev.pol.boot.n.content.B100000.sspro.p1.rcr-vs-generic-ppi-source.reject\endcsname{100.0}
\expandafter\def\csname ppt@rev.pol.boot.n.content.B100000.sspro.p1.rcr-vs-generic-ppi-source.reject\endcsname{$+$100.0}
\expandafter\def\csname val@rev.pol.boot.n.content.B100000.sspro.p1.rcr-vs-replay-ppi-uniform.p-holm\endcsname{0.036}
\expandafter\def\csname cnt@rev.pol.boot.n.content.B100000.sspro.p1.rcr-vs-replay-ppi-uniform.p-holm\endcsname{0}
\expandafter\def\csname pct@rev.pol.boot.n.content.B100000.sspro.p1.rcr-vs-replay-ppi-uniform.p-holm\endcsname{3.6}
\expandafter\def\csname ppt@rev.pol.boot.n.content.B100000.sspro.p1.rcr-vs-replay-ppi-uniform.p-holm\endcsname{$+$3.6}
\expandafter\def\csname val@rev.pol.boot.n.content.B100000.sspro.p1.rcr-vs-replay-ppi-uniform.p-raw\endcsname{0.009}
\expandafter\def\csname cnt@rev.pol.boot.n.content.B100000.sspro.p1.rcr-vs-replay-ppi-uniform.p-raw\endcsname{0}
\expandafter\def\csname pct@rev.pol.boot.n.content.B100000.sspro.p1.rcr-vs-replay-ppi-uniform.p-raw\endcsname{0.9}
\expandafter\def\csname ppt@rev.pol.boot.n.content.B100000.sspro.p1.rcr-vs-replay-ppi-uniform.p-raw\endcsname{$+$0.9}
\expandafter\def\csname val@rev.pol.boot.n.content.B100000.sspro.p1.rcr-vs-replay-ppi-uniform.reject\endcsname{1}
\expandafter\def\csname cnt@rev.pol.boot.n.content.B100000.sspro.p1.rcr-vs-replay-ppi-uniform.reject\endcsname{1}
\expandafter\def\csname pct@rev.pol.boot.n.content.B100000.sspro.p1.rcr-vs-replay-ppi-uniform.reject\endcsname{100.0}
\expandafter\def\csname ppt@rev.pol.boot.n.content.B100000.sspro.p1.rcr-vs-replay-ppi-uniform.reject\endcsname{$+$100.0}
\expandafter\def\csname val@rev.pol.boot.n.content.B100000.sspro.p2.rcr-vs-generic-ppi-source.p-holm\endcsname{0.889}
\expandafter\def\csname cnt@rev.pol.boot.n.content.B100000.sspro.p2.rcr-vs-generic-ppi-source.p-holm\endcsname{1}
\expandafter\def\csname pct@rev.pol.boot.n.content.B100000.sspro.p2.rcr-vs-generic-ppi-source.p-holm\endcsname{88.9}
\expandafter\def\csname ppt@rev.pol.boot.n.content.B100000.sspro.p2.rcr-vs-generic-ppi-source.p-holm\endcsname{$+$88.9}
\expandafter\def\csname val@rev.pol.boot.n.content.B100000.sspro.p2.rcr-vs-generic-ppi-source.p-raw\endcsname{0.585}
\expandafter\def\csname cnt@rev.pol.boot.n.content.B100000.sspro.p2.rcr-vs-generic-ppi-source.p-raw\endcsname{1}
\expandafter\def\csname pct@rev.pol.boot.n.content.B100000.sspro.p2.rcr-vs-generic-ppi-source.p-raw\endcsname{58.5}
\expandafter\def\csname ppt@rev.pol.boot.n.content.B100000.sspro.p2.rcr-vs-generic-ppi-source.p-raw\endcsname{$+$58.5}
\expandafter\def\csname val@rev.pol.boot.n.content.B100000.sspro.p2.rcr-vs-generic-ppi-source.reject\endcsname{0}
\expandafter\def\csname cnt@rev.pol.boot.n.content.B100000.sspro.p2.rcr-vs-generic-ppi-source.reject\endcsname{0}
\expandafter\def\csname pct@rev.pol.boot.n.content.B100000.sspro.p2.rcr-vs-generic-ppi-source.reject\endcsname{0.0}
\expandafter\def\csname ppt@rev.pol.boot.n.content.B100000.sspro.p2.rcr-vs-generic-ppi-source.reject\endcsname{0.0}
\expandafter\def\csname val@rev.pol.boot.n.content.B100000.sspro.p2.rcr-vs-replay-ppi-uniform.p-holm\endcsname{0.889}
\expandafter\def\csname cnt@rev.pol.boot.n.content.B100000.sspro.p2.rcr-vs-replay-ppi-uniform.p-holm\endcsname{1}
\expandafter\def\csname pct@rev.pol.boot.n.content.B100000.sspro.p2.rcr-vs-replay-ppi-uniform.p-holm\endcsname{88.9}
\expandafter\def\csname ppt@rev.pol.boot.n.content.B100000.sspro.p2.rcr-vs-replay-ppi-uniform.p-holm\endcsname{$+$88.9}
\expandafter\def\csname val@rev.pol.boot.n.content.B100000.sspro.p2.rcr-vs-replay-ppi-uniform.p-raw\endcsname{0.445}
\expandafter\def\csname cnt@rev.pol.boot.n.content.B100000.sspro.p2.rcr-vs-replay-ppi-uniform.p-raw\endcsname{0}
\expandafter\def\csname pct@rev.pol.boot.n.content.B100000.sspro.p2.rcr-vs-replay-ppi-uniform.p-raw\endcsname{44.5}
\expandafter\def\csname ppt@rev.pol.boot.n.content.B100000.sspro.p2.rcr-vs-replay-ppi-uniform.p-raw\endcsname{$+$44.5}
\expandafter\def\csname val@rev.pol.boot.n.content.B100000.sspro.p2.rcr-vs-replay-ppi-uniform.reject\endcsname{0}
\expandafter\def\csname cnt@rev.pol.boot.n.content.B100000.sspro.p2.rcr-vs-replay-ppi-uniform.reject\endcsname{0}
\expandafter\def\csname pct@rev.pol.boot.n.content.B100000.sspro.p2.rcr-vs-replay-ppi-uniform.reject\endcsname{0.0}
\expandafter\def\csname ppt@rev.pol.boot.n.content.B100000.sspro.p2.rcr-vs-replay-ppi-uniform.reject\endcsname{0.0}
\expandafter\def\csname val@rev.pol.boot.n.content.B100000.uiv.p1.rcr-vs-generic-ppi-source.p-holm\endcsname{0.000}
\expandafter\def\csname cnt@rev.pol.boot.n.content.B100000.uiv.p1.rcr-vs-generic-ppi-source.p-holm\endcsname{0}
\expandafter\def\csname pct@rev.pol.boot.n.content.B100000.uiv.p1.rcr-vs-generic-ppi-source.p-holm\endcsname{0.0}
\expandafter\def\csname ppt@rev.pol.boot.n.content.B100000.uiv.p1.rcr-vs-generic-ppi-source.p-holm\endcsname{0.0}
\expandafter\def\csname val@rev.pol.boot.n.content.B100000.uiv.p1.rcr-vs-generic-ppi-source.p-raw\endcsname{0.000}
\expandafter\def\csname cnt@rev.pol.boot.n.content.B100000.uiv.p1.rcr-vs-generic-ppi-source.p-raw\endcsname{0}
\expandafter\def\csname pct@rev.pol.boot.n.content.B100000.uiv.p1.rcr-vs-generic-ppi-source.p-raw\endcsname{0.0}
\expandafter\def\csname ppt@rev.pol.boot.n.content.B100000.uiv.p1.rcr-vs-generic-ppi-source.p-raw\endcsname{0.0}
\expandafter\def\csname val@rev.pol.boot.n.content.B100000.uiv.p1.rcr-vs-generic-ppi-source.reject\endcsname{1}
\expandafter\def\csname cnt@rev.pol.boot.n.content.B100000.uiv.p1.rcr-vs-generic-ppi-source.reject\endcsname{1}
\expandafter\def\csname pct@rev.pol.boot.n.content.B100000.uiv.p1.rcr-vs-generic-ppi-source.reject\endcsname{100.0}
\expandafter\def\csname ppt@rev.pol.boot.n.content.B100000.uiv.p1.rcr-vs-generic-ppi-source.reject\endcsname{$+$100.0}
\expandafter\def\csname val@rev.pol.boot.n.content.B100000.uiv.p1.rcr-vs-replay-ppi-uniform.p-holm\endcsname{0.000}
\expandafter\def\csname cnt@rev.pol.boot.n.content.B100000.uiv.p1.rcr-vs-replay-ppi-uniform.p-holm\endcsname{0}
\expandafter\def\csname pct@rev.pol.boot.n.content.B100000.uiv.p1.rcr-vs-replay-ppi-uniform.p-holm\endcsname{0.0}
\expandafter\def\csname ppt@rev.pol.boot.n.content.B100000.uiv.p1.rcr-vs-replay-ppi-uniform.p-holm\endcsname{0.0}
\expandafter\def\csname val@rev.pol.boot.n.content.B100000.uiv.p1.rcr-vs-replay-ppi-uniform.p-raw\endcsname{0.000}
\expandafter\def\csname cnt@rev.pol.boot.n.content.B100000.uiv.p1.rcr-vs-replay-ppi-uniform.p-raw\endcsname{0}
\expandafter\def\csname pct@rev.pol.boot.n.content.B100000.uiv.p1.rcr-vs-replay-ppi-uniform.p-raw\endcsname{0.0}
\expandafter\def\csname ppt@rev.pol.boot.n.content.B100000.uiv.p1.rcr-vs-replay-ppi-uniform.p-raw\endcsname{0.0}
\expandafter\def\csname val@rev.pol.boot.n.content.B100000.uiv.p1.rcr-vs-replay-ppi-uniform.reject\endcsname{1}
\expandafter\def\csname cnt@rev.pol.boot.n.content.B100000.uiv.p1.rcr-vs-replay-ppi-uniform.reject\endcsname{1}
\expandafter\def\csname pct@rev.pol.boot.n.content.B100000.uiv.p1.rcr-vs-replay-ppi-uniform.reject\endcsname{100.0}
\expandafter\def\csname ppt@rev.pol.boot.n.content.B100000.uiv.p1.rcr-vs-replay-ppi-uniform.reject\endcsname{$+$100.0}
\expandafter\def\csname val@rev.pol.boot.n.content.B100000.uiv.p2.rcr-vs-generic-ppi-source.p-holm\endcsname{0.134}
\expandafter\def\csname cnt@rev.pol.boot.n.content.B100000.uiv.p2.rcr-vs-generic-ppi-source.p-holm\endcsname{0}
\expandafter\def\csname pct@rev.pol.boot.n.content.B100000.uiv.p2.rcr-vs-generic-ppi-source.p-holm\endcsname{13.4}
\expandafter\def\csname ppt@rev.pol.boot.n.content.B100000.uiv.p2.rcr-vs-generic-ppi-source.p-holm\endcsname{$+$13.4}
\expandafter\def\csname val@rev.pol.boot.n.content.B100000.uiv.p2.rcr-vs-generic-ppi-source.p-raw\endcsname{0.045}
\expandafter\def\csname cnt@rev.pol.boot.n.content.B100000.uiv.p2.rcr-vs-generic-ppi-source.p-raw\endcsname{0}
\expandafter\def\csname pct@rev.pol.boot.n.content.B100000.uiv.p2.rcr-vs-generic-ppi-source.p-raw\endcsname{4.5}
\expandafter\def\csname ppt@rev.pol.boot.n.content.B100000.uiv.p2.rcr-vs-generic-ppi-source.p-raw\endcsname{$+$4.5}
\expandafter\def\csname val@rev.pol.boot.n.content.B100000.uiv.p2.rcr-vs-generic-ppi-source.reject\endcsname{0}
\expandafter\def\csname cnt@rev.pol.boot.n.content.B100000.uiv.p2.rcr-vs-generic-ppi-source.reject\endcsname{0}
\expandafter\def\csname pct@rev.pol.boot.n.content.B100000.uiv.p2.rcr-vs-generic-ppi-source.reject\endcsname{0.0}
\expandafter\def\csname ppt@rev.pol.boot.n.content.B100000.uiv.p2.rcr-vs-generic-ppi-source.reject\endcsname{0.0}
\expandafter\def\csname val@rev.pol.boot.n.content.B100000.uiv.p2.rcr-vs-replay-ppi-uniform.p-holm\endcsname{0.019}
\expandafter\def\csname cnt@rev.pol.boot.n.content.B100000.uiv.p2.rcr-vs-replay-ppi-uniform.p-holm\endcsname{0}
\expandafter\def\csname pct@rev.pol.boot.n.content.B100000.uiv.p2.rcr-vs-replay-ppi-uniform.p-holm\endcsname{1.9}
\expandafter\def\csname ppt@rev.pol.boot.n.content.B100000.uiv.p2.rcr-vs-replay-ppi-uniform.p-holm\endcsname{$+$1.9}
\expandafter\def\csname val@rev.pol.boot.n.content.B100000.uiv.p2.rcr-vs-replay-ppi-uniform.p-raw\endcsname{0.003}
\expandafter\def\csname cnt@rev.pol.boot.n.content.B100000.uiv.p2.rcr-vs-replay-ppi-uniform.p-raw\endcsname{0}
\expandafter\def\csname pct@rev.pol.boot.n.content.B100000.uiv.p2.rcr-vs-replay-ppi-uniform.p-raw\endcsname{0.3}
\expandafter\def\csname ppt@rev.pol.boot.n.content.B100000.uiv.p2.rcr-vs-replay-ppi-uniform.p-raw\endcsname{$+$0.3}
\expandafter\def\csname val@rev.pol.boot.n.content.B100000.uiv.p2.rcr-vs-replay-ppi-uniform.reject\endcsname{1}
\expandafter\def\csname cnt@rev.pol.boot.n.content.B100000.uiv.p2.rcr-vs-replay-ppi-uniform.reject\endcsname{1}
\expandafter\def\csname pct@rev.pol.boot.n.content.B100000.uiv.p2.rcr-vs-replay-ppi-uniform.reject\endcsname{100.0}
\expandafter\def\csname ppt@rev.pol.boot.n.content.B100000.uiv.p2.rcr-vs-replay-ppi-uniform.reject\endcsname{$+$100.0}
\expandafter\def\csname val@rev.pol.boot.n.content.B4000.n-rejected\endcsname{5}
\expandafter\def\csname cnt@rev.pol.boot.n.content.B4000.n-rejected\endcsname{5}
\expandafter\def\csname pct@rev.pol.boot.n.content.B4000.n-rejected\endcsname{500.0}
\expandafter\def\csname ppt@rev.pol.boot.n.content.B4000.n-rejected\endcsname{$+$500.0}
\expandafter\def\csname val@rev.pol.boot.n.content.B4000.sspro.p1.rcr-vs-generic-ppi-source.p-holm\endcsname{0.024}
\expandafter\def\csname cnt@rev.pol.boot.n.content.B4000.sspro.p1.rcr-vs-generic-ppi-source.p-holm\endcsname{0}
\expandafter\def\csname pct@rev.pol.boot.n.content.B4000.sspro.p1.rcr-vs-generic-ppi-source.p-holm\endcsname{2.4}
\expandafter\def\csname ppt@rev.pol.boot.n.content.B4000.sspro.p1.rcr-vs-generic-ppi-source.p-holm\endcsname{$+$2.4}
\expandafter\def\csname val@rev.pol.boot.n.content.B4000.sspro.p1.rcr-vs-generic-ppi-source.p-raw\endcsname{0.005}
\expandafter\def\csname cnt@rev.pol.boot.n.content.B4000.sspro.p1.rcr-vs-generic-ppi-source.p-raw\endcsname{0}
\expandafter\def\csname pct@rev.pol.boot.n.content.B4000.sspro.p1.rcr-vs-generic-ppi-source.p-raw\endcsname{0.5}
\expandafter\def\csname ppt@rev.pol.boot.n.content.B4000.sspro.p1.rcr-vs-generic-ppi-source.p-raw\endcsname{$+$0.5}
\expandafter\def\csname val@rev.pol.boot.n.content.B4000.sspro.p1.rcr-vs-generic-ppi-source.reject\endcsname{1}
\expandafter\def\csname cnt@rev.pol.boot.n.content.B4000.sspro.p1.rcr-vs-generic-ppi-source.reject\endcsname{1}
\expandafter\def\csname pct@rev.pol.boot.n.content.B4000.sspro.p1.rcr-vs-generic-ppi-source.reject\endcsname{100.0}
\expandafter\def\csname ppt@rev.pol.boot.n.content.B4000.sspro.p1.rcr-vs-generic-ppi-source.reject\endcsname{$+$100.0}
\expandafter\def\csname val@rev.pol.boot.n.content.B4000.sspro.p1.rcr-vs-replay-ppi-uniform.p-holm\endcsname{0.035}
\expandafter\def\csname cnt@rev.pol.boot.n.content.B4000.sspro.p1.rcr-vs-replay-ppi-uniform.p-holm\endcsname{0}
\expandafter\def\csname pct@rev.pol.boot.n.content.B4000.sspro.p1.rcr-vs-replay-ppi-uniform.p-holm\endcsname{3.5}
\expandafter\def\csname ppt@rev.pol.boot.n.content.B4000.sspro.p1.rcr-vs-replay-ppi-uniform.p-holm\endcsname{$+$3.5}
\expandafter\def\csname val@rev.pol.boot.n.content.B4000.sspro.p1.rcr-vs-replay-ppi-uniform.p-raw\endcsname{0.009}
\expandafter\def\csname cnt@rev.pol.boot.n.content.B4000.sspro.p1.rcr-vs-replay-ppi-uniform.p-raw\endcsname{0}
\expandafter\def\csname pct@rev.pol.boot.n.content.B4000.sspro.p1.rcr-vs-replay-ppi-uniform.p-raw\endcsname{0.9}
\expandafter\def\csname ppt@rev.pol.boot.n.content.B4000.sspro.p1.rcr-vs-replay-ppi-uniform.p-raw\endcsname{$+$0.9}
\expandafter\def\csname val@rev.pol.boot.n.content.B4000.sspro.p1.rcr-vs-replay-ppi-uniform.reject\endcsname{1}
\expandafter\def\csname cnt@rev.pol.boot.n.content.B4000.sspro.p1.rcr-vs-replay-ppi-uniform.reject\endcsname{1}
\expandafter\def\csname pct@rev.pol.boot.n.content.B4000.sspro.p1.rcr-vs-replay-ppi-uniform.reject\endcsname{100.0}
\expandafter\def\csname ppt@rev.pol.boot.n.content.B4000.sspro.p1.rcr-vs-replay-ppi-uniform.reject\endcsname{$+$100.0}
\expandafter\def\csname val@rev.pol.boot.n.content.B4000.sspro.p2.rcr-vs-generic-ppi-source.p-holm\endcsname{0.876}
\expandafter\def\csname cnt@rev.pol.boot.n.content.B4000.sspro.p2.rcr-vs-generic-ppi-source.p-holm\endcsname{1}
\expandafter\def\csname pct@rev.pol.boot.n.content.B4000.sspro.p2.rcr-vs-generic-ppi-source.p-holm\endcsname{87.6}
\expandafter\def\csname ppt@rev.pol.boot.n.content.B4000.sspro.p2.rcr-vs-generic-ppi-source.p-holm\endcsname{$+$87.6}
\expandafter\def\csname val@rev.pol.boot.n.content.B4000.sspro.p2.rcr-vs-generic-ppi-source.p-raw\endcsname{0.577}
\expandafter\def\csname cnt@rev.pol.boot.n.content.B4000.sspro.p2.rcr-vs-generic-ppi-source.p-raw\endcsname{1}
\expandafter\def\csname pct@rev.pol.boot.n.content.B4000.sspro.p2.rcr-vs-generic-ppi-source.p-raw\endcsname{57.7}
\expandafter\def\csname ppt@rev.pol.boot.n.content.B4000.sspro.p2.rcr-vs-generic-ppi-source.p-raw\endcsname{$+$57.7}
\expandafter\def\csname val@rev.pol.boot.n.content.B4000.sspro.p2.rcr-vs-generic-ppi-source.reject\endcsname{0}
\expandafter\def\csname cnt@rev.pol.boot.n.content.B4000.sspro.p2.rcr-vs-generic-ppi-source.reject\endcsname{0}
\expandafter\def\csname pct@rev.pol.boot.n.content.B4000.sspro.p2.rcr-vs-generic-ppi-source.reject\endcsname{0.0}
\expandafter\def\csname ppt@rev.pol.boot.n.content.B4000.sspro.p2.rcr-vs-generic-ppi-source.reject\endcsname{0.0}
\expandafter\def\csname val@rev.pol.boot.n.content.B4000.sspro.p2.rcr-vs-replay-ppi-uniform.p-holm\endcsname{0.876}
\expandafter\def\csname cnt@rev.pol.boot.n.content.B4000.sspro.p2.rcr-vs-replay-ppi-uniform.p-holm\endcsname{1}
\expandafter\def\csname pct@rev.pol.boot.n.content.B4000.sspro.p2.rcr-vs-replay-ppi-uniform.p-holm\endcsname{87.6}
\expandafter\def\csname ppt@rev.pol.boot.n.content.B4000.sspro.p2.rcr-vs-replay-ppi-uniform.p-holm\endcsname{$+$87.6}
\expandafter\def\csname val@rev.pol.boot.n.content.B4000.sspro.p2.rcr-vs-replay-ppi-uniform.p-raw\endcsname{0.438}
\expandafter\def\csname cnt@rev.pol.boot.n.content.B4000.sspro.p2.rcr-vs-replay-ppi-uniform.p-raw\endcsname{0}
\expandafter\def\csname pct@rev.pol.boot.n.content.B4000.sspro.p2.rcr-vs-replay-ppi-uniform.p-raw\endcsname{43.8}
\expandafter\def\csname ppt@rev.pol.boot.n.content.B4000.sspro.p2.rcr-vs-replay-ppi-uniform.p-raw\endcsname{$+$43.8}
\expandafter\def\csname val@rev.pol.boot.n.content.B4000.sspro.p2.rcr-vs-replay-ppi-uniform.reject\endcsname{0}
\expandafter\def\csname cnt@rev.pol.boot.n.content.B4000.sspro.p2.rcr-vs-replay-ppi-uniform.reject\endcsname{0}
\expandafter\def\csname pct@rev.pol.boot.n.content.B4000.sspro.p2.rcr-vs-replay-ppi-uniform.reject\endcsname{0.0}
\expandafter\def\csname ppt@rev.pol.boot.n.content.B4000.sspro.p2.rcr-vs-replay-ppi-uniform.reject\endcsname{0.0}
\expandafter\def\csname val@rev.pol.boot.n.content.B4000.uiv.p1.rcr-vs-generic-ppi-source.p-holm\endcsname{0.002}
\expandafter\def\csname cnt@rev.pol.boot.n.content.B4000.uiv.p1.rcr-vs-generic-ppi-source.p-holm\endcsname{0}
\expandafter\def\csname pct@rev.pol.boot.n.content.B4000.uiv.p1.rcr-vs-generic-ppi-source.p-holm\endcsname{0.2}
\expandafter\def\csname ppt@rev.pol.boot.n.content.B4000.uiv.p1.rcr-vs-generic-ppi-source.p-holm\endcsname{$+$0.2}
\expandafter\def\csname val@rev.pol.boot.n.content.B4000.uiv.p1.rcr-vs-generic-ppi-source.p-raw\endcsname{0.000}
\expandafter\def\csname cnt@rev.pol.boot.n.content.B4000.uiv.p1.rcr-vs-generic-ppi-source.p-raw\endcsname{0}
\expandafter\def\csname pct@rev.pol.boot.n.content.B4000.uiv.p1.rcr-vs-generic-ppi-source.p-raw\endcsname{0.0}
\expandafter\def\csname ppt@rev.pol.boot.n.content.B4000.uiv.p1.rcr-vs-generic-ppi-source.p-raw\endcsname{0.0}
\expandafter\def\csname val@rev.pol.boot.n.content.B4000.uiv.p1.rcr-vs-generic-ppi-source.reject\endcsname{1}
\expandafter\def\csname cnt@rev.pol.boot.n.content.B4000.uiv.p1.rcr-vs-generic-ppi-source.reject\endcsname{1}
\expandafter\def\csname pct@rev.pol.boot.n.content.B4000.uiv.p1.rcr-vs-generic-ppi-source.reject\endcsname{100.0}
\expandafter\def\csname ppt@rev.pol.boot.n.content.B4000.uiv.p1.rcr-vs-generic-ppi-source.reject\endcsname{$+$100.0}
\expandafter\def\csname val@rev.pol.boot.n.content.B4000.uiv.p1.rcr-vs-replay-ppi-uniform.p-holm\endcsname{0.002}
\expandafter\def\csname cnt@rev.pol.boot.n.content.B4000.uiv.p1.rcr-vs-replay-ppi-uniform.p-holm\endcsname{0}
\expandafter\def\csname pct@rev.pol.boot.n.content.B4000.uiv.p1.rcr-vs-replay-ppi-uniform.p-holm\endcsname{0.2}
\expandafter\def\csname ppt@rev.pol.boot.n.content.B4000.uiv.p1.rcr-vs-replay-ppi-uniform.p-holm\endcsname{$+$0.2}
\expandafter\def\csname val@rev.pol.boot.n.content.B4000.uiv.p1.rcr-vs-replay-ppi-uniform.p-raw\endcsname{0.000}
\expandafter\def\csname cnt@rev.pol.boot.n.content.B4000.uiv.p1.rcr-vs-replay-ppi-uniform.p-raw\endcsname{0}
\expandafter\def\csname pct@rev.pol.boot.n.content.B4000.uiv.p1.rcr-vs-replay-ppi-uniform.p-raw\endcsname{0.0}
\expandafter\def\csname ppt@rev.pol.boot.n.content.B4000.uiv.p1.rcr-vs-replay-ppi-uniform.p-raw\endcsname{0.0}
\expandafter\def\csname val@rev.pol.boot.n.content.B4000.uiv.p1.rcr-vs-replay-ppi-uniform.reject\endcsname{1}
\expandafter\def\csname cnt@rev.pol.boot.n.content.B4000.uiv.p1.rcr-vs-replay-ppi-uniform.reject\endcsname{1}
\expandafter\def\csname pct@rev.pol.boot.n.content.B4000.uiv.p1.rcr-vs-replay-ppi-uniform.reject\endcsname{100.0}
\expandafter\def\csname ppt@rev.pol.boot.n.content.B4000.uiv.p1.rcr-vs-replay-ppi-uniform.reject\endcsname{$+$100.0}
\expandafter\def\csname val@rev.pol.boot.n.content.B4000.uiv.p2.rcr-vs-generic-ppi-source.p-holm\endcsname{0.137}
\expandafter\def\csname cnt@rev.pol.boot.n.content.B4000.uiv.p2.rcr-vs-generic-ppi-source.p-holm\endcsname{0}
\expandafter\def\csname pct@rev.pol.boot.n.content.B4000.uiv.p2.rcr-vs-generic-ppi-source.p-holm\endcsname{13.7}
\expandafter\def\csname ppt@rev.pol.boot.n.content.B4000.uiv.p2.rcr-vs-generic-ppi-source.p-holm\endcsname{$+$13.7}
\expandafter\def\csname val@rev.pol.boot.n.content.B4000.uiv.p2.rcr-vs-generic-ppi-source.p-raw\endcsname{0.046}
\expandafter\def\csname cnt@rev.pol.boot.n.content.B4000.uiv.p2.rcr-vs-generic-ppi-source.p-raw\endcsname{0}
\expandafter\def\csname pct@rev.pol.boot.n.content.B4000.uiv.p2.rcr-vs-generic-ppi-source.p-raw\endcsname{4.6}
\expandafter\def\csname ppt@rev.pol.boot.n.content.B4000.uiv.p2.rcr-vs-generic-ppi-source.p-raw\endcsname{$+$4.6}
\expandafter\def\csname val@rev.pol.boot.n.content.B4000.uiv.p2.rcr-vs-generic-ppi-source.reject\endcsname{0}
\expandafter\def\csname cnt@rev.pol.boot.n.content.B4000.uiv.p2.rcr-vs-generic-ppi-source.reject\endcsname{0}
\expandafter\def\csname pct@rev.pol.boot.n.content.B4000.uiv.p2.rcr-vs-generic-ppi-source.reject\endcsname{0.0}
\expandafter\def\csname ppt@rev.pol.boot.n.content.B4000.uiv.p2.rcr-vs-generic-ppi-source.reject\endcsname{0.0}
\expandafter\def\csname val@rev.pol.boot.n.content.B4000.uiv.p2.rcr-vs-replay-ppi-uniform.p-holm\endcsname{0.021}
\expandafter\def\csname cnt@rev.pol.boot.n.content.B4000.uiv.p2.rcr-vs-replay-ppi-uniform.p-holm\endcsname{0}
\expandafter\def\csname pct@rev.pol.boot.n.content.B4000.uiv.p2.rcr-vs-replay-ppi-uniform.p-holm\endcsname{2.1}
\expandafter\def\csname ppt@rev.pol.boot.n.content.B4000.uiv.p2.rcr-vs-replay-ppi-uniform.p-holm\endcsname{$+$2.1}
\expandafter\def\csname val@rev.pol.boot.n.content.B4000.uiv.p2.rcr-vs-replay-ppi-uniform.p-raw\endcsname{0.003}
\expandafter\def\csname cnt@rev.pol.boot.n.content.B4000.uiv.p2.rcr-vs-replay-ppi-uniform.p-raw\endcsname{0}
\expandafter\def\csname pct@rev.pol.boot.n.content.B4000.uiv.p2.rcr-vs-replay-ppi-uniform.p-raw\endcsname{0.3}
\expandafter\def\csname ppt@rev.pol.boot.n.content.B4000.uiv.p2.rcr-vs-replay-ppi-uniform.p-raw\endcsname{$+$0.3}
\expandafter\def\csname val@rev.pol.boot.n.content.B4000.uiv.p2.rcr-vs-replay-ppi-uniform.reject\endcsname{1}
\expandafter\def\csname cnt@rev.pol.boot.n.content.B4000.uiv.p2.rcr-vs-replay-ppi-uniform.reject\endcsname{1}
\expandafter\def\csname pct@rev.pol.boot.n.content.B4000.uiv.p2.rcr-vs-replay-ppi-uniform.reject\endcsname{100.0}
\expandafter\def\csname ppt@rev.pol.boot.n.content.B4000.uiv.p2.rcr-vs-replay-ppi-uniform.reject\endcsname{$+$100.0}
\expandafter\def\csname val@rev.pol.boot.n.fixedA.B100000.max-p-holm-rejected\endcsname{0.036}
\expandafter\def\csname cnt@rev.pol.boot.n.fixedA.B100000.max-p-holm-rejected\endcsname{0}
\expandafter\def\csname pct@rev.pol.boot.n.fixedA.B100000.max-p-holm-rejected\endcsname{3.6}
\expandafter\def\csname ppt@rev.pol.boot.n.fixedA.B100000.max-p-holm-rejected\endcsname{$+$3.6}
\expandafter\def\csname val@rev.pol.boot.n.fixedA.B100000.n-rejected\endcsname{5}
\expandafter\def\csname cnt@rev.pol.boot.n.fixedA.B100000.n-rejected\endcsname{5}
\expandafter\def\csname pct@rev.pol.boot.n.fixedA.B100000.n-rejected\endcsname{500.0}
\expandafter\def\csname ppt@rev.pol.boot.n.fixedA.B100000.n-rejected\endcsname{$+$500.0}
\expandafter\def\csname val@rev.pol.boot.n.fixedA.B100000.sspro.p1.rcr-vs-generic-ppi-source.p-holm\endcsname{0.027}
\expandafter\def\csname cnt@rev.pol.boot.n.fixedA.B100000.sspro.p1.rcr-vs-generic-ppi-source.p-holm\endcsname{0}
\expandafter\def\csname pct@rev.pol.boot.n.fixedA.B100000.sspro.p1.rcr-vs-generic-ppi-source.p-holm\endcsname{2.7}
\expandafter\def\csname ppt@rev.pol.boot.n.fixedA.B100000.sspro.p1.rcr-vs-generic-ppi-source.p-holm\endcsname{$+$2.7}
\expandafter\def\csname val@rev.pol.boot.n.fixedA.B100000.sspro.p1.rcr-vs-generic-ppi-source.p-raw\endcsname{0.005}
\expandafter\def\csname cnt@rev.pol.boot.n.fixedA.B100000.sspro.p1.rcr-vs-generic-ppi-source.p-raw\endcsname{0}
\expandafter\def\csname pct@rev.pol.boot.n.fixedA.B100000.sspro.p1.rcr-vs-generic-ppi-source.p-raw\endcsname{0.5}
\expandafter\def\csname ppt@rev.pol.boot.n.fixedA.B100000.sspro.p1.rcr-vs-generic-ppi-source.p-raw\endcsname{$+$0.5}
\expandafter\def\csname val@rev.pol.boot.n.fixedA.B100000.sspro.p1.rcr-vs-generic-ppi-source.reject\endcsname{1}
\expandafter\def\csname cnt@rev.pol.boot.n.fixedA.B100000.sspro.p1.rcr-vs-generic-ppi-source.reject\endcsname{1}
\expandafter\def\csname pct@rev.pol.boot.n.fixedA.B100000.sspro.p1.rcr-vs-generic-ppi-source.reject\endcsname{100.0}
\expandafter\def\csname ppt@rev.pol.boot.n.fixedA.B100000.sspro.p1.rcr-vs-generic-ppi-source.reject\endcsname{$+$100.0}
\expandafter\def\csname val@rev.pol.boot.n.fixedA.B100000.sspro.p1.rcr-vs-replay-ppi-uniform.p-holm\endcsname{0.036}
\expandafter\def\csname cnt@rev.pol.boot.n.fixedA.B100000.sspro.p1.rcr-vs-replay-ppi-uniform.p-holm\endcsname{0}
\expandafter\def\csname pct@rev.pol.boot.n.fixedA.B100000.sspro.p1.rcr-vs-replay-ppi-uniform.p-holm\endcsname{3.6}
\expandafter\def\csname ppt@rev.pol.boot.n.fixedA.B100000.sspro.p1.rcr-vs-replay-ppi-uniform.p-holm\endcsname{$+$3.6}
\expandafter\def\csname val@rev.pol.boot.n.fixedA.B100000.sspro.p1.rcr-vs-replay-ppi-uniform.p-raw\endcsname{0.009}
\expandafter\def\csname cnt@rev.pol.boot.n.fixedA.B100000.sspro.p1.rcr-vs-replay-ppi-uniform.p-raw\endcsname{0}
\expandafter\def\csname pct@rev.pol.boot.n.fixedA.B100000.sspro.p1.rcr-vs-replay-ppi-uniform.p-raw\endcsname{0.9}
\expandafter\def\csname ppt@rev.pol.boot.n.fixedA.B100000.sspro.p1.rcr-vs-replay-ppi-uniform.p-raw\endcsname{$+$0.9}
\expandafter\def\csname val@rev.pol.boot.n.fixedA.B100000.sspro.p1.rcr-vs-replay-ppi-uniform.reject\endcsname{1}
\expandafter\def\csname cnt@rev.pol.boot.n.fixedA.B100000.sspro.p1.rcr-vs-replay-ppi-uniform.reject\endcsname{1}
\expandafter\def\csname pct@rev.pol.boot.n.fixedA.B100000.sspro.p1.rcr-vs-replay-ppi-uniform.reject\endcsname{100.0}
\expandafter\def\csname ppt@rev.pol.boot.n.fixedA.B100000.sspro.p1.rcr-vs-replay-ppi-uniform.reject\endcsname{$+$100.0}
\expandafter\def\csname val@rev.pol.boot.n.fixedA.B100000.sspro.p2.rcr-vs-generic-ppi-source.p-holm\endcsname{0.891}
\expandafter\def\csname cnt@rev.pol.boot.n.fixedA.B100000.sspro.p2.rcr-vs-generic-ppi-source.p-holm\endcsname{1}
\expandafter\def\csname pct@rev.pol.boot.n.fixedA.B100000.sspro.p2.rcr-vs-generic-ppi-source.p-holm\endcsname{89.1}
\expandafter\def\csname ppt@rev.pol.boot.n.fixedA.B100000.sspro.p2.rcr-vs-generic-ppi-source.p-holm\endcsname{$+$89.1}
\expandafter\def\csname val@rev.pol.boot.n.fixedA.B100000.sspro.p2.rcr-vs-generic-ppi-source.p-raw\endcsname{0.585}
\expandafter\def\csname cnt@rev.pol.boot.n.fixedA.B100000.sspro.p2.rcr-vs-generic-ppi-source.p-raw\endcsname{1}
\expandafter\def\csname pct@rev.pol.boot.n.fixedA.B100000.sspro.p2.rcr-vs-generic-ppi-source.p-raw\endcsname{58.5}
\expandafter\def\csname ppt@rev.pol.boot.n.fixedA.B100000.sspro.p2.rcr-vs-generic-ppi-source.p-raw\endcsname{$+$58.5}
\expandafter\def\csname val@rev.pol.boot.n.fixedA.B100000.sspro.p2.rcr-vs-generic-ppi-source.reject\endcsname{0}
\expandafter\def\csname cnt@rev.pol.boot.n.fixedA.B100000.sspro.p2.rcr-vs-generic-ppi-source.reject\endcsname{0}
\expandafter\def\csname pct@rev.pol.boot.n.fixedA.B100000.sspro.p2.rcr-vs-generic-ppi-source.reject\endcsname{0.0}
\expandafter\def\csname ppt@rev.pol.boot.n.fixedA.B100000.sspro.p2.rcr-vs-generic-ppi-source.reject\endcsname{0.0}
\expandafter\def\csname val@rev.pol.boot.n.fixedA.B100000.sspro.p2.rcr-vs-replay-ppi-uniform.p-holm\endcsname{0.891}
\expandafter\def\csname cnt@rev.pol.boot.n.fixedA.B100000.sspro.p2.rcr-vs-replay-ppi-uniform.p-holm\endcsname{1}
\expandafter\def\csname pct@rev.pol.boot.n.fixedA.B100000.sspro.p2.rcr-vs-replay-ppi-uniform.p-holm\endcsname{89.1}
\expandafter\def\csname ppt@rev.pol.boot.n.fixedA.B100000.sspro.p2.rcr-vs-replay-ppi-uniform.p-holm\endcsname{$+$89.1}
\expandafter\def\csname val@rev.pol.boot.n.fixedA.B100000.sspro.p2.rcr-vs-replay-ppi-uniform.p-raw\endcsname{0.446}
\expandafter\def\csname cnt@rev.pol.boot.n.fixedA.B100000.sspro.p2.rcr-vs-replay-ppi-uniform.p-raw\endcsname{0}
\expandafter\def\csname pct@rev.pol.boot.n.fixedA.B100000.sspro.p2.rcr-vs-replay-ppi-uniform.p-raw\endcsname{44.6}
\expandafter\def\csname ppt@rev.pol.boot.n.fixedA.B100000.sspro.p2.rcr-vs-replay-ppi-uniform.p-raw\endcsname{$+$44.6}
\expandafter\def\csname val@rev.pol.boot.n.fixedA.B100000.sspro.p2.rcr-vs-replay-ppi-uniform.reject\endcsname{0}
\expandafter\def\csname cnt@rev.pol.boot.n.fixedA.B100000.sspro.p2.rcr-vs-replay-ppi-uniform.reject\endcsname{0}
\expandafter\def\csname pct@rev.pol.boot.n.fixedA.B100000.sspro.p2.rcr-vs-replay-ppi-uniform.reject\endcsname{0.0}
\expandafter\def\csname ppt@rev.pol.boot.n.fixedA.B100000.sspro.p2.rcr-vs-replay-ppi-uniform.reject\endcsname{0.0}
\expandafter\def\csname val@rev.pol.boot.n.fixedA.B100000.uiv.p1.rcr-vs-generic-ppi-source.p-holm\endcsname{0.001}
\expandafter\def\csname cnt@rev.pol.boot.n.fixedA.B100000.uiv.p1.rcr-vs-generic-ppi-source.p-holm\endcsname{0}
\expandafter\def\csname pct@rev.pol.boot.n.fixedA.B100000.uiv.p1.rcr-vs-generic-ppi-source.p-holm\endcsname{0.1}
\expandafter\def\csname ppt@rev.pol.boot.n.fixedA.B100000.uiv.p1.rcr-vs-generic-ppi-source.p-holm\endcsname{$+$0.1}
\expandafter\def\csname val@rev.pol.boot.n.fixedA.B100000.uiv.p1.rcr-vs-generic-ppi-source.p-raw\endcsname{0.000}
\expandafter\def\csname cnt@rev.pol.boot.n.fixedA.B100000.uiv.p1.rcr-vs-generic-ppi-source.p-raw\endcsname{0}
\expandafter\def\csname pct@rev.pol.boot.n.fixedA.B100000.uiv.p1.rcr-vs-generic-ppi-source.p-raw\endcsname{0.0}
\expandafter\def\csname ppt@rev.pol.boot.n.fixedA.B100000.uiv.p1.rcr-vs-generic-ppi-source.p-raw\endcsname{0.0}
\expandafter\def\csname val@rev.pol.boot.n.fixedA.B100000.uiv.p1.rcr-vs-generic-ppi-source.reject\endcsname{1}
\expandafter\def\csname cnt@rev.pol.boot.n.fixedA.B100000.uiv.p1.rcr-vs-generic-ppi-source.reject\endcsname{1}
\expandafter\def\csname pct@rev.pol.boot.n.fixedA.B100000.uiv.p1.rcr-vs-generic-ppi-source.reject\endcsname{100.0}
\expandafter\def\csname ppt@rev.pol.boot.n.fixedA.B100000.uiv.p1.rcr-vs-generic-ppi-source.reject\endcsname{$+$100.0}
\expandafter\def\csname val@rev.pol.boot.n.fixedA.B100000.uiv.p1.rcr-vs-replay-ppi-uniform.p-holm\endcsname{0.000}
\expandafter\def\csname cnt@rev.pol.boot.n.fixedA.B100000.uiv.p1.rcr-vs-replay-ppi-uniform.p-holm\endcsname{0}
\expandafter\def\csname pct@rev.pol.boot.n.fixedA.B100000.uiv.p1.rcr-vs-replay-ppi-uniform.p-holm\endcsname{0.0}
\expandafter\def\csname ppt@rev.pol.boot.n.fixedA.B100000.uiv.p1.rcr-vs-replay-ppi-uniform.p-holm\endcsname{0.0}
\expandafter\def\csname val@rev.pol.boot.n.fixedA.B100000.uiv.p1.rcr-vs-replay-ppi-uniform.p-raw\endcsname{0.000}
\expandafter\def\csname cnt@rev.pol.boot.n.fixedA.B100000.uiv.p1.rcr-vs-replay-ppi-uniform.p-raw\endcsname{0}
\expandafter\def\csname pct@rev.pol.boot.n.fixedA.B100000.uiv.p1.rcr-vs-replay-ppi-uniform.p-raw\endcsname{0.0}
\expandafter\def\csname ppt@rev.pol.boot.n.fixedA.B100000.uiv.p1.rcr-vs-replay-ppi-uniform.p-raw\endcsname{0.0}
\expandafter\def\csname val@rev.pol.boot.n.fixedA.B100000.uiv.p1.rcr-vs-replay-ppi-uniform.reject\endcsname{1}
\expandafter\def\csname cnt@rev.pol.boot.n.fixedA.B100000.uiv.p1.rcr-vs-replay-ppi-uniform.reject\endcsname{1}
\expandafter\def\csname pct@rev.pol.boot.n.fixedA.B100000.uiv.p1.rcr-vs-replay-ppi-uniform.reject\endcsname{100.0}
\expandafter\def\csname ppt@rev.pol.boot.n.fixedA.B100000.uiv.p1.rcr-vs-replay-ppi-uniform.reject\endcsname{$+$100.0}
\expandafter\def\csname val@rev.pol.boot.n.fixedA.B100000.uiv.p2.rcr-vs-generic-ppi-source.p-holm\endcsname{0.134}
\expandafter\def\csname cnt@rev.pol.boot.n.fixedA.B100000.uiv.p2.rcr-vs-generic-ppi-source.p-holm\endcsname{0}
\expandafter\def\csname pct@rev.pol.boot.n.fixedA.B100000.uiv.p2.rcr-vs-generic-ppi-source.p-holm\endcsname{13.4}
\expandafter\def\csname ppt@rev.pol.boot.n.fixedA.B100000.uiv.p2.rcr-vs-generic-ppi-source.p-holm\endcsname{$+$13.4}
\expandafter\def\csname val@rev.pol.boot.n.fixedA.B100000.uiv.p2.rcr-vs-generic-ppi-source.p-raw\endcsname{0.045}
\expandafter\def\csname cnt@rev.pol.boot.n.fixedA.B100000.uiv.p2.rcr-vs-generic-ppi-source.p-raw\endcsname{0}
\expandafter\def\csname pct@rev.pol.boot.n.fixedA.B100000.uiv.p2.rcr-vs-generic-ppi-source.p-raw\endcsname{4.5}
\expandafter\def\csname ppt@rev.pol.boot.n.fixedA.B100000.uiv.p2.rcr-vs-generic-ppi-source.p-raw\endcsname{$+$4.5}
\expandafter\def\csname val@rev.pol.boot.n.fixedA.B100000.uiv.p2.rcr-vs-generic-ppi-source.reject\endcsname{0}
\expandafter\def\csname cnt@rev.pol.boot.n.fixedA.B100000.uiv.p2.rcr-vs-generic-ppi-source.reject\endcsname{0}
\expandafter\def\csname pct@rev.pol.boot.n.fixedA.B100000.uiv.p2.rcr-vs-generic-ppi-source.reject\endcsname{0.0}
\expandafter\def\csname ppt@rev.pol.boot.n.fixedA.B100000.uiv.p2.rcr-vs-generic-ppi-source.reject\endcsname{0.0}
\expandafter\def\csname val@rev.pol.boot.n.fixedA.B100000.uiv.p2.rcr-vs-replay-ppi-uniform.p-holm\endcsname{0.020}
\expandafter\def\csname cnt@rev.pol.boot.n.fixedA.B100000.uiv.p2.rcr-vs-replay-ppi-uniform.p-holm\endcsname{0}
\expandafter\def\csname pct@rev.pol.boot.n.fixedA.B100000.uiv.p2.rcr-vs-replay-ppi-uniform.p-holm\endcsname{2.0}
\expandafter\def\csname ppt@rev.pol.boot.n.fixedA.B100000.uiv.p2.rcr-vs-replay-ppi-uniform.p-holm\endcsname{$+$2.0}
\expandafter\def\csname val@rev.pol.boot.n.fixedA.B100000.uiv.p2.rcr-vs-replay-ppi-uniform.p-raw\endcsname{0.003}
\expandafter\def\csname cnt@rev.pol.boot.n.fixedA.B100000.uiv.p2.rcr-vs-replay-ppi-uniform.p-raw\endcsname{0}
\expandafter\def\csname pct@rev.pol.boot.n.fixedA.B100000.uiv.p2.rcr-vs-replay-ppi-uniform.p-raw\endcsname{0.3}
\expandafter\def\csname ppt@rev.pol.boot.n.fixedA.B100000.uiv.p2.rcr-vs-replay-ppi-uniform.p-raw\endcsname{$+$0.3}
\expandafter\def\csname val@rev.pol.boot.n.fixedA.B100000.uiv.p2.rcr-vs-replay-ppi-uniform.reject\endcsname{1}
\expandafter\def\csname cnt@rev.pol.boot.n.fixedA.B100000.uiv.p2.rcr-vs-replay-ppi-uniform.reject\endcsname{1}
\expandafter\def\csname pct@rev.pol.boot.n.fixedA.B100000.uiv.p2.rcr-vs-replay-ppi-uniform.reject\endcsname{100.0}
\expandafter\def\csname ppt@rev.pol.boot.n.fixedA.B100000.uiv.p2.rcr-vs-replay-ppi-uniform.reject\endcsname{$+$100.0}
\expandafter\def\csname val@rev.pol.boot.n.fixedA.B4000.n-rejected\endcsname{5}
\expandafter\def\csname cnt@rev.pol.boot.n.fixedA.B4000.n-rejected\endcsname{5}
\expandafter\def\csname pct@rev.pol.boot.n.fixedA.B4000.n-rejected\endcsname{500.0}
\expandafter\def\csname ppt@rev.pol.boot.n.fixedA.B4000.n-rejected\endcsname{$+$500.0}
\expandafter\def\csname val@rev.pol.boot.n.fixedA.B4000.sspro.p1.rcr-vs-generic-ppi-source.p-holm\endcsname{0.024}
\expandafter\def\csname cnt@rev.pol.boot.n.fixedA.B4000.sspro.p1.rcr-vs-generic-ppi-source.p-holm\endcsname{0}
\expandafter\def\csname pct@rev.pol.boot.n.fixedA.B4000.sspro.p1.rcr-vs-generic-ppi-source.p-holm\endcsname{2.4}
\expandafter\def\csname ppt@rev.pol.boot.n.fixedA.B4000.sspro.p1.rcr-vs-generic-ppi-source.p-holm\endcsname{$+$2.4}
\expandafter\def\csname val@rev.pol.boot.n.fixedA.B4000.sspro.p1.rcr-vs-generic-ppi-source.p-raw\endcsname{0.004}
\expandafter\def\csname cnt@rev.pol.boot.n.fixedA.B4000.sspro.p1.rcr-vs-generic-ppi-source.p-raw\endcsname{0}
\expandafter\def\csname pct@rev.pol.boot.n.fixedA.B4000.sspro.p1.rcr-vs-generic-ppi-source.p-raw\endcsname{0.4}
\expandafter\def\csname ppt@rev.pol.boot.n.fixedA.B4000.sspro.p1.rcr-vs-generic-ppi-source.p-raw\endcsname{$+$0.4}
\expandafter\def\csname val@rev.pol.boot.n.fixedA.B4000.sspro.p1.rcr-vs-generic-ppi-source.reject\endcsname{1}
\expandafter\def\csname cnt@rev.pol.boot.n.fixedA.B4000.sspro.p1.rcr-vs-generic-ppi-source.reject\endcsname{1}
\expandafter\def\csname pct@rev.pol.boot.n.fixedA.B4000.sspro.p1.rcr-vs-generic-ppi-source.reject\endcsname{100.0}
\expandafter\def\csname ppt@rev.pol.boot.n.fixedA.B4000.sspro.p1.rcr-vs-generic-ppi-source.reject\endcsname{$+$100.0}
\expandafter\def\csname val@rev.pol.boot.n.fixedA.B4000.sspro.p1.rcr-vs-replay-ppi-uniform.p-holm\endcsname{0.030}
\expandafter\def\csname cnt@rev.pol.boot.n.fixedA.B4000.sspro.p1.rcr-vs-replay-ppi-uniform.p-holm\endcsname{0}
\expandafter\def\csname pct@rev.pol.boot.n.fixedA.B4000.sspro.p1.rcr-vs-replay-ppi-uniform.p-holm\endcsname{3.0}
\expandafter\def\csname ppt@rev.pol.boot.n.fixedA.B4000.sspro.p1.rcr-vs-replay-ppi-uniform.p-holm\endcsname{$+$3.0}
\expandafter\def\csname val@rev.pol.boot.n.fixedA.B4000.sspro.p1.rcr-vs-replay-ppi-uniform.p-raw\endcsname{0.007}
\expandafter\def\csname cnt@rev.pol.boot.n.fixedA.B4000.sspro.p1.rcr-vs-replay-ppi-uniform.p-raw\endcsname{0}
\expandafter\def\csname pct@rev.pol.boot.n.fixedA.B4000.sspro.p1.rcr-vs-replay-ppi-uniform.p-raw\endcsname{0.7}
\expandafter\def\csname ppt@rev.pol.boot.n.fixedA.B4000.sspro.p1.rcr-vs-replay-ppi-uniform.p-raw\endcsname{$+$0.7}
\expandafter\def\csname val@rev.pol.boot.n.fixedA.B4000.sspro.p1.rcr-vs-replay-ppi-uniform.reject\endcsname{1}
\expandafter\def\csname cnt@rev.pol.boot.n.fixedA.B4000.sspro.p1.rcr-vs-replay-ppi-uniform.reject\endcsname{1}
\expandafter\def\csname pct@rev.pol.boot.n.fixedA.B4000.sspro.p1.rcr-vs-replay-ppi-uniform.reject\endcsname{100.0}
\expandafter\def\csname ppt@rev.pol.boot.n.fixedA.B4000.sspro.p1.rcr-vs-replay-ppi-uniform.reject\endcsname{$+$100.0}
\expandafter\def\csname val@rev.pol.boot.n.fixedA.B4000.sspro.p2.rcr-vs-generic-ppi-source.p-holm\endcsname{0.907}
\expandafter\def\csname cnt@rev.pol.boot.n.fixedA.B4000.sspro.p2.rcr-vs-generic-ppi-source.p-holm\endcsname{1}
\expandafter\def\csname pct@rev.pol.boot.n.fixedA.B4000.sspro.p2.rcr-vs-generic-ppi-source.p-holm\endcsname{90.7}
\expandafter\def\csname ppt@rev.pol.boot.n.fixedA.B4000.sspro.p2.rcr-vs-generic-ppi-source.p-holm\endcsname{$+$90.7}
\expandafter\def\csname val@rev.pol.boot.n.fixedA.B4000.sspro.p2.rcr-vs-generic-ppi-source.p-raw\endcsname{0.581}
\expandafter\def\csname cnt@rev.pol.boot.n.fixedA.B4000.sspro.p2.rcr-vs-generic-ppi-source.p-raw\endcsname{1}
\expandafter\def\csname pct@rev.pol.boot.n.fixedA.B4000.sspro.p2.rcr-vs-generic-ppi-source.p-raw\endcsname{58.1}
\expandafter\def\csname ppt@rev.pol.boot.n.fixedA.B4000.sspro.p2.rcr-vs-generic-ppi-source.p-raw\endcsname{$+$58.1}
\expandafter\def\csname val@rev.pol.boot.n.fixedA.B4000.sspro.p2.rcr-vs-generic-ppi-source.reject\endcsname{0}
\expandafter\def\csname cnt@rev.pol.boot.n.fixedA.B4000.sspro.p2.rcr-vs-generic-ppi-source.reject\endcsname{0}
\expandafter\def\csname pct@rev.pol.boot.n.fixedA.B4000.sspro.p2.rcr-vs-generic-ppi-source.reject\endcsname{0.0}
\expandafter\def\csname ppt@rev.pol.boot.n.fixedA.B4000.sspro.p2.rcr-vs-generic-ppi-source.reject\endcsname{0.0}
\expandafter\def\csname val@rev.pol.boot.n.fixedA.B4000.sspro.p2.rcr-vs-replay-ppi-uniform.p-holm\endcsname{0.907}
\expandafter\def\csname cnt@rev.pol.boot.n.fixedA.B4000.sspro.p2.rcr-vs-replay-ppi-uniform.p-holm\endcsname{1}
\expandafter\def\csname pct@rev.pol.boot.n.fixedA.B4000.sspro.p2.rcr-vs-replay-ppi-uniform.p-holm\endcsname{90.7}
\expandafter\def\csname ppt@rev.pol.boot.n.fixedA.B4000.sspro.p2.rcr-vs-replay-ppi-uniform.p-holm\endcsname{$+$90.7}
\expandafter\def\csname val@rev.pol.boot.n.fixedA.B4000.sspro.p2.rcr-vs-replay-ppi-uniform.p-raw\endcsname{0.454}
\expandafter\def\csname cnt@rev.pol.boot.n.fixedA.B4000.sspro.p2.rcr-vs-replay-ppi-uniform.p-raw\endcsname{0}
\expandafter\def\csname pct@rev.pol.boot.n.fixedA.B4000.sspro.p2.rcr-vs-replay-ppi-uniform.p-raw\endcsname{45.4}
\expandafter\def\csname ppt@rev.pol.boot.n.fixedA.B4000.sspro.p2.rcr-vs-replay-ppi-uniform.p-raw\endcsname{$+$45.4}
\expandafter\def\csname val@rev.pol.boot.n.fixedA.B4000.sspro.p2.rcr-vs-replay-ppi-uniform.reject\endcsname{0}
\expandafter\def\csname cnt@rev.pol.boot.n.fixedA.B4000.sspro.p2.rcr-vs-replay-ppi-uniform.reject\endcsname{0}
\expandafter\def\csname pct@rev.pol.boot.n.fixedA.B4000.sspro.p2.rcr-vs-replay-ppi-uniform.reject\endcsname{0.0}
\expandafter\def\csname ppt@rev.pol.boot.n.fixedA.B4000.sspro.p2.rcr-vs-replay-ppi-uniform.reject\endcsname{0.0}
\expandafter\def\csname val@rev.pol.boot.n.fixedA.B4000.uiv.p1.rcr-vs-generic-ppi-source.p-holm\endcsname{0.002}
\expandafter\def\csname cnt@rev.pol.boot.n.fixedA.B4000.uiv.p1.rcr-vs-generic-ppi-source.p-holm\endcsname{0}
\expandafter\def\csname pct@rev.pol.boot.n.fixedA.B4000.uiv.p1.rcr-vs-generic-ppi-source.p-holm\endcsname{0.2}
\expandafter\def\csname ppt@rev.pol.boot.n.fixedA.B4000.uiv.p1.rcr-vs-generic-ppi-source.p-holm\endcsname{$+$0.2}
\expandafter\def\csname val@rev.pol.boot.n.fixedA.B4000.uiv.p1.rcr-vs-generic-ppi-source.p-raw\endcsname{0.000}
\expandafter\def\csname cnt@rev.pol.boot.n.fixedA.B4000.uiv.p1.rcr-vs-generic-ppi-source.p-raw\endcsname{0}
\expandafter\def\csname pct@rev.pol.boot.n.fixedA.B4000.uiv.p1.rcr-vs-generic-ppi-source.p-raw\endcsname{0.0}
\expandafter\def\csname ppt@rev.pol.boot.n.fixedA.B4000.uiv.p1.rcr-vs-generic-ppi-source.p-raw\endcsname{0.0}
\expandafter\def\csname val@rev.pol.boot.n.fixedA.B4000.uiv.p1.rcr-vs-generic-ppi-source.reject\endcsname{1}
\expandafter\def\csname cnt@rev.pol.boot.n.fixedA.B4000.uiv.p1.rcr-vs-generic-ppi-source.reject\endcsname{1}
\expandafter\def\csname pct@rev.pol.boot.n.fixedA.B4000.uiv.p1.rcr-vs-generic-ppi-source.reject\endcsname{100.0}
\expandafter\def\csname ppt@rev.pol.boot.n.fixedA.B4000.uiv.p1.rcr-vs-generic-ppi-source.reject\endcsname{$+$100.0}
\expandafter\def\csname val@rev.pol.boot.n.fixedA.B4000.uiv.p1.rcr-vs-replay-ppi-uniform.p-holm\endcsname{0.003}
\expandafter\def\csname cnt@rev.pol.boot.n.fixedA.B4000.uiv.p1.rcr-vs-replay-ppi-uniform.p-holm\endcsname{0}
\expandafter\def\csname pct@rev.pol.boot.n.fixedA.B4000.uiv.p1.rcr-vs-replay-ppi-uniform.p-holm\endcsname{0.3}
\expandafter\def\csname ppt@rev.pol.boot.n.fixedA.B4000.uiv.p1.rcr-vs-replay-ppi-uniform.p-holm\endcsname{$+$0.3}
\expandafter\def\csname val@rev.pol.boot.n.fixedA.B4000.uiv.p1.rcr-vs-replay-ppi-uniform.p-raw\endcsname{0.000}
\expandafter\def\csname cnt@rev.pol.boot.n.fixedA.B4000.uiv.p1.rcr-vs-replay-ppi-uniform.p-raw\endcsname{0}
\expandafter\def\csname pct@rev.pol.boot.n.fixedA.B4000.uiv.p1.rcr-vs-replay-ppi-uniform.p-raw\endcsname{0.0}
\expandafter\def\csname ppt@rev.pol.boot.n.fixedA.B4000.uiv.p1.rcr-vs-replay-ppi-uniform.p-raw\endcsname{0.0}
\expandafter\def\csname val@rev.pol.boot.n.fixedA.B4000.uiv.p1.rcr-vs-replay-ppi-uniform.reject\endcsname{1}
\expandafter\def\csname cnt@rev.pol.boot.n.fixedA.B4000.uiv.p1.rcr-vs-replay-ppi-uniform.reject\endcsname{1}
\expandafter\def\csname pct@rev.pol.boot.n.fixedA.B4000.uiv.p1.rcr-vs-replay-ppi-uniform.reject\endcsname{100.0}
\expandafter\def\csname ppt@rev.pol.boot.n.fixedA.B4000.uiv.p1.rcr-vs-replay-ppi-uniform.reject\endcsname{$+$100.0}
\expandafter\def\csname val@rev.pol.boot.n.fixedA.B4000.uiv.p2.rcr-vs-generic-ppi-source.p-holm\endcsname{0.135}
\expandafter\def\csname cnt@rev.pol.boot.n.fixedA.B4000.uiv.p2.rcr-vs-generic-ppi-source.p-holm\endcsname{0}
\expandafter\def\csname pct@rev.pol.boot.n.fixedA.B4000.uiv.p2.rcr-vs-generic-ppi-source.p-holm\endcsname{13.5}
\expandafter\def\csname ppt@rev.pol.boot.n.fixedA.B4000.uiv.p2.rcr-vs-generic-ppi-source.p-holm\endcsname{$+$13.5}
\expandafter\def\csname val@rev.pol.boot.n.fixedA.B4000.uiv.p2.rcr-vs-generic-ppi-source.p-raw\endcsname{0.045}
\expandafter\def\csname cnt@rev.pol.boot.n.fixedA.B4000.uiv.p2.rcr-vs-generic-ppi-source.p-raw\endcsname{0}
\expandafter\def\csname pct@rev.pol.boot.n.fixedA.B4000.uiv.p2.rcr-vs-generic-ppi-source.p-raw\endcsname{4.5}
\expandafter\def\csname ppt@rev.pol.boot.n.fixedA.B4000.uiv.p2.rcr-vs-generic-ppi-source.p-raw\endcsname{$+$4.5}
\expandafter\def\csname val@rev.pol.boot.n.fixedA.B4000.uiv.p2.rcr-vs-generic-ppi-source.reject\endcsname{0}
\expandafter\def\csname cnt@rev.pol.boot.n.fixedA.B4000.uiv.p2.rcr-vs-generic-ppi-source.reject\endcsname{0}
\expandafter\def\csname pct@rev.pol.boot.n.fixedA.B4000.uiv.p2.rcr-vs-generic-ppi-source.reject\endcsname{0.0}
\expandafter\def\csname ppt@rev.pol.boot.n.fixedA.B4000.uiv.p2.rcr-vs-generic-ppi-source.reject\endcsname{0.0}
\expandafter\def\csname val@rev.pol.boot.n.fixedA.B4000.uiv.p2.rcr-vs-replay-ppi-uniform.p-holm\endcsname{0.024}
\expandafter\def\csname cnt@rev.pol.boot.n.fixedA.B4000.uiv.p2.rcr-vs-replay-ppi-uniform.p-holm\endcsname{0}
\expandafter\def\csname pct@rev.pol.boot.n.fixedA.B4000.uiv.p2.rcr-vs-replay-ppi-uniform.p-holm\endcsname{2.4}
\expandafter\def\csname ppt@rev.pol.boot.n.fixedA.B4000.uiv.p2.rcr-vs-replay-ppi-uniform.p-holm\endcsname{$+$2.4}
\expandafter\def\csname val@rev.pol.boot.n.fixedA.B4000.uiv.p2.rcr-vs-replay-ppi-uniform.p-raw\endcsname{0.004}
\expandafter\def\csname cnt@rev.pol.boot.n.fixedA.B4000.uiv.p2.rcr-vs-replay-ppi-uniform.p-raw\endcsname{0}
\expandafter\def\csname pct@rev.pol.boot.n.fixedA.B4000.uiv.p2.rcr-vs-replay-ppi-uniform.p-raw\endcsname{0.4}
\expandafter\def\csname ppt@rev.pol.boot.n.fixedA.B4000.uiv.p2.rcr-vs-replay-ppi-uniform.p-raw\endcsname{$+$0.4}
\expandafter\def\csname val@rev.pol.boot.n.fixedA.B4000.uiv.p2.rcr-vs-replay-ppi-uniform.reject\endcsname{1}
\expandafter\def\csname cnt@rev.pol.boot.n.fixedA.B4000.uiv.p2.rcr-vs-replay-ppi-uniform.reject\endcsname{1}
\expandafter\def\csname pct@rev.pol.boot.n.fixedA.B4000.uiv.p2.rcr-vs-replay-ppi-uniform.reject\endcsname{100.0}
\expandafter\def\csname ppt@rev.pol.boot.n.fixedA.B4000.uiv.p2.rcr-vs-replay-ppi-uniform.reject\endcsname{$+$100.0}
\expandafter\def\csname val@rev.pol.boot.n.fixedB.B100000.max-p-holm-rejected\endcsname{0.037}
\expandafter\def\csname cnt@rev.pol.boot.n.fixedB.B100000.max-p-holm-rejected\endcsname{0}
\expandafter\def\csname pct@rev.pol.boot.n.fixedB.B100000.max-p-holm-rejected\endcsname{3.7}
\expandafter\def\csname ppt@rev.pol.boot.n.fixedB.B100000.max-p-holm-rejected\endcsname{$+$3.7}
\expandafter\def\csname val@rev.pol.boot.n.fixedB.B100000.n-rejected\endcsname{5}
\expandafter\def\csname cnt@rev.pol.boot.n.fixedB.B100000.n-rejected\endcsname{5}
\expandafter\def\csname pct@rev.pol.boot.n.fixedB.B100000.n-rejected\endcsname{500.0}
\expandafter\def\csname ppt@rev.pol.boot.n.fixedB.B100000.n-rejected\endcsname{$+$500.0}
\expandafter\def\csname val@rev.pol.boot.n.fixedB.B100000.sspro.p1.rcr-vs-generic-ppi-source.p-holm\endcsname{0.028}
\expandafter\def\csname cnt@rev.pol.boot.n.fixedB.B100000.sspro.p1.rcr-vs-generic-ppi-source.p-holm\endcsname{0}
\expandafter\def\csname pct@rev.pol.boot.n.fixedB.B100000.sspro.p1.rcr-vs-generic-ppi-source.p-holm\endcsname{2.8}
\expandafter\def\csname ppt@rev.pol.boot.n.fixedB.B100000.sspro.p1.rcr-vs-generic-ppi-source.p-holm\endcsname{$+$2.8}
\expandafter\def\csname val@rev.pol.boot.n.fixedB.B100000.sspro.p1.rcr-vs-generic-ppi-source.p-raw\endcsname{0.006}
\expandafter\def\csname cnt@rev.pol.boot.n.fixedB.B100000.sspro.p1.rcr-vs-generic-ppi-source.p-raw\endcsname{0}
\expandafter\def\csname pct@rev.pol.boot.n.fixedB.B100000.sspro.p1.rcr-vs-generic-ppi-source.p-raw\endcsname{0.6}
\expandafter\def\csname ppt@rev.pol.boot.n.fixedB.B100000.sspro.p1.rcr-vs-generic-ppi-source.p-raw\endcsname{$+$0.6}
\expandafter\def\csname val@rev.pol.boot.n.fixedB.B100000.sspro.p1.rcr-vs-generic-ppi-source.reject\endcsname{1}
\expandafter\def\csname cnt@rev.pol.boot.n.fixedB.B100000.sspro.p1.rcr-vs-generic-ppi-source.reject\endcsname{1}
\expandafter\def\csname pct@rev.pol.boot.n.fixedB.B100000.sspro.p1.rcr-vs-generic-ppi-source.reject\endcsname{100.0}
\expandafter\def\csname ppt@rev.pol.boot.n.fixedB.B100000.sspro.p1.rcr-vs-generic-ppi-source.reject\endcsname{$+$100.0}
\expandafter\def\csname val@rev.pol.boot.n.fixedB.B100000.sspro.p1.rcr-vs-replay-ppi-uniform.p-holm\endcsname{0.037}
\expandafter\def\csname cnt@rev.pol.boot.n.fixedB.B100000.sspro.p1.rcr-vs-replay-ppi-uniform.p-holm\endcsname{0}
\expandafter\def\csname pct@rev.pol.boot.n.fixedB.B100000.sspro.p1.rcr-vs-replay-ppi-uniform.p-holm\endcsname{3.7}
\expandafter\def\csname ppt@rev.pol.boot.n.fixedB.B100000.sspro.p1.rcr-vs-replay-ppi-uniform.p-holm\endcsname{$+$3.7}
\expandafter\def\csname val@rev.pol.boot.n.fixedB.B100000.sspro.p1.rcr-vs-replay-ppi-uniform.p-raw\endcsname{0.009}
\expandafter\def\csname cnt@rev.pol.boot.n.fixedB.B100000.sspro.p1.rcr-vs-replay-ppi-uniform.p-raw\endcsname{0}
\expandafter\def\csname pct@rev.pol.boot.n.fixedB.B100000.sspro.p1.rcr-vs-replay-ppi-uniform.p-raw\endcsname{0.9}
\expandafter\def\csname ppt@rev.pol.boot.n.fixedB.B100000.sspro.p1.rcr-vs-replay-ppi-uniform.p-raw\endcsname{$+$0.9}
\expandafter\def\csname val@rev.pol.boot.n.fixedB.B100000.sspro.p1.rcr-vs-replay-ppi-uniform.reject\endcsname{1}
\expandafter\def\csname cnt@rev.pol.boot.n.fixedB.B100000.sspro.p1.rcr-vs-replay-ppi-uniform.reject\endcsname{1}
\expandafter\def\csname pct@rev.pol.boot.n.fixedB.B100000.sspro.p1.rcr-vs-replay-ppi-uniform.reject\endcsname{100.0}
\expandafter\def\csname ppt@rev.pol.boot.n.fixedB.B100000.sspro.p1.rcr-vs-replay-ppi-uniform.reject\endcsname{$+$100.0}
\expandafter\def\csname val@rev.pol.boot.n.fixedB.B100000.sspro.p2.rcr-vs-generic-ppi-source.p-holm\endcsname{0.893}
\expandafter\def\csname cnt@rev.pol.boot.n.fixedB.B100000.sspro.p2.rcr-vs-generic-ppi-source.p-holm\endcsname{1}
\expandafter\def\csname pct@rev.pol.boot.n.fixedB.B100000.sspro.p2.rcr-vs-generic-ppi-source.p-holm\endcsname{89.3}
\expandafter\def\csname ppt@rev.pol.boot.n.fixedB.B100000.sspro.p2.rcr-vs-generic-ppi-source.p-holm\endcsname{$+$89.3}
\expandafter\def\csname val@rev.pol.boot.n.fixedB.B100000.sspro.p2.rcr-vs-generic-ppi-source.p-raw\endcsname{0.584}
\expandafter\def\csname cnt@rev.pol.boot.n.fixedB.B100000.sspro.p2.rcr-vs-generic-ppi-source.p-raw\endcsname{1}
\expandafter\def\csname pct@rev.pol.boot.n.fixedB.B100000.sspro.p2.rcr-vs-generic-ppi-source.p-raw\endcsname{58.4}
\expandafter\def\csname ppt@rev.pol.boot.n.fixedB.B100000.sspro.p2.rcr-vs-generic-ppi-source.p-raw\endcsname{$+$58.4}
\expandafter\def\csname val@rev.pol.boot.n.fixedB.B100000.sspro.p2.rcr-vs-generic-ppi-source.reject\endcsname{0}
\expandafter\def\csname cnt@rev.pol.boot.n.fixedB.B100000.sspro.p2.rcr-vs-generic-ppi-source.reject\endcsname{0}
\expandafter\def\csname pct@rev.pol.boot.n.fixedB.B100000.sspro.p2.rcr-vs-generic-ppi-source.reject\endcsname{0.0}
\expandafter\def\csname ppt@rev.pol.boot.n.fixedB.B100000.sspro.p2.rcr-vs-generic-ppi-source.reject\endcsname{0.0}
\expandafter\def\csname val@rev.pol.boot.n.fixedB.B100000.sspro.p2.rcr-vs-replay-ppi-uniform.p-holm\endcsname{0.893}
\expandafter\def\csname cnt@rev.pol.boot.n.fixedB.B100000.sspro.p2.rcr-vs-replay-ppi-uniform.p-holm\endcsname{1}
\expandafter\def\csname pct@rev.pol.boot.n.fixedB.B100000.sspro.p2.rcr-vs-replay-ppi-uniform.p-holm\endcsname{89.3}
\expandafter\def\csname ppt@rev.pol.boot.n.fixedB.B100000.sspro.p2.rcr-vs-replay-ppi-uniform.p-holm\endcsname{$+$89.3}
\expandafter\def\csname val@rev.pol.boot.n.fixedB.B100000.sspro.p2.rcr-vs-replay-ppi-uniform.p-raw\endcsname{0.446}
\expandafter\def\csname cnt@rev.pol.boot.n.fixedB.B100000.sspro.p2.rcr-vs-replay-ppi-uniform.p-raw\endcsname{0}
\expandafter\def\csname pct@rev.pol.boot.n.fixedB.B100000.sspro.p2.rcr-vs-replay-ppi-uniform.p-raw\endcsname{44.6}
\expandafter\def\csname ppt@rev.pol.boot.n.fixedB.B100000.sspro.p2.rcr-vs-replay-ppi-uniform.p-raw\endcsname{$+$44.6}
\expandafter\def\csname val@rev.pol.boot.n.fixedB.B100000.sspro.p2.rcr-vs-replay-ppi-uniform.reject\endcsname{0}
\expandafter\def\csname cnt@rev.pol.boot.n.fixedB.B100000.sspro.p2.rcr-vs-replay-ppi-uniform.reject\endcsname{0}
\expandafter\def\csname pct@rev.pol.boot.n.fixedB.B100000.sspro.p2.rcr-vs-replay-ppi-uniform.reject\endcsname{0.0}
\expandafter\def\csname ppt@rev.pol.boot.n.fixedB.B100000.sspro.p2.rcr-vs-replay-ppi-uniform.reject\endcsname{0.0}
\expandafter\def\csname val@rev.pol.boot.n.fixedB.B100000.uiv.p1.rcr-vs-generic-ppi-source.p-holm\endcsname{0.001}
\expandafter\def\csname cnt@rev.pol.boot.n.fixedB.B100000.uiv.p1.rcr-vs-generic-ppi-source.p-holm\endcsname{0}
\expandafter\def\csname pct@rev.pol.boot.n.fixedB.B100000.uiv.p1.rcr-vs-generic-ppi-source.p-holm\endcsname{0.1}
\expandafter\def\csname ppt@rev.pol.boot.n.fixedB.B100000.uiv.p1.rcr-vs-generic-ppi-source.p-holm\endcsname{$+$0.1}
\expandafter\def\csname val@rev.pol.boot.n.fixedB.B100000.uiv.p1.rcr-vs-generic-ppi-source.p-raw\endcsname{0.000}
\expandafter\def\csname cnt@rev.pol.boot.n.fixedB.B100000.uiv.p1.rcr-vs-generic-ppi-source.p-raw\endcsname{0}
\expandafter\def\csname pct@rev.pol.boot.n.fixedB.B100000.uiv.p1.rcr-vs-generic-ppi-source.p-raw\endcsname{0.0}
\expandafter\def\csname ppt@rev.pol.boot.n.fixedB.B100000.uiv.p1.rcr-vs-generic-ppi-source.p-raw\endcsname{0.0}
\expandafter\def\csname val@rev.pol.boot.n.fixedB.B100000.uiv.p1.rcr-vs-generic-ppi-source.reject\endcsname{1}
\expandafter\def\csname cnt@rev.pol.boot.n.fixedB.B100000.uiv.p1.rcr-vs-generic-ppi-source.reject\endcsname{1}
\expandafter\def\csname pct@rev.pol.boot.n.fixedB.B100000.uiv.p1.rcr-vs-generic-ppi-source.reject\endcsname{100.0}
\expandafter\def\csname ppt@rev.pol.boot.n.fixedB.B100000.uiv.p1.rcr-vs-generic-ppi-source.reject\endcsname{$+$100.0}
\expandafter\def\csname val@rev.pol.boot.n.fixedB.B100000.uiv.p1.rcr-vs-replay-ppi-uniform.p-holm\endcsname{0.000}
\expandafter\def\csname cnt@rev.pol.boot.n.fixedB.B100000.uiv.p1.rcr-vs-replay-ppi-uniform.p-holm\endcsname{0}
\expandafter\def\csname pct@rev.pol.boot.n.fixedB.B100000.uiv.p1.rcr-vs-replay-ppi-uniform.p-holm\endcsname{0.0}
\expandafter\def\csname ppt@rev.pol.boot.n.fixedB.B100000.uiv.p1.rcr-vs-replay-ppi-uniform.p-holm\endcsname{0.0}
\expandafter\def\csname val@rev.pol.boot.n.fixedB.B100000.uiv.p1.rcr-vs-replay-ppi-uniform.p-raw\endcsname{0.000}
\expandafter\def\csname cnt@rev.pol.boot.n.fixedB.B100000.uiv.p1.rcr-vs-replay-ppi-uniform.p-raw\endcsname{0}
\expandafter\def\csname pct@rev.pol.boot.n.fixedB.B100000.uiv.p1.rcr-vs-replay-ppi-uniform.p-raw\endcsname{0.0}
\expandafter\def\csname ppt@rev.pol.boot.n.fixedB.B100000.uiv.p1.rcr-vs-replay-ppi-uniform.p-raw\endcsname{0.0}
\expandafter\def\csname val@rev.pol.boot.n.fixedB.B100000.uiv.p1.rcr-vs-replay-ppi-uniform.reject\endcsname{1}
\expandafter\def\csname cnt@rev.pol.boot.n.fixedB.B100000.uiv.p1.rcr-vs-replay-ppi-uniform.reject\endcsname{1}
\expandafter\def\csname pct@rev.pol.boot.n.fixedB.B100000.uiv.p1.rcr-vs-replay-ppi-uniform.reject\endcsname{100.0}
\expandafter\def\csname ppt@rev.pol.boot.n.fixedB.B100000.uiv.p1.rcr-vs-replay-ppi-uniform.reject\endcsname{$+$100.0}
\expandafter\def\csname val@rev.pol.boot.n.fixedB.B100000.uiv.p2.rcr-vs-generic-ppi-source.p-holm\endcsname{0.130}
\expandafter\def\csname cnt@rev.pol.boot.n.fixedB.B100000.uiv.p2.rcr-vs-generic-ppi-source.p-holm\endcsname{0}
\expandafter\def\csname pct@rev.pol.boot.n.fixedB.B100000.uiv.p2.rcr-vs-generic-ppi-source.p-holm\endcsname{13.0}
\expandafter\def\csname ppt@rev.pol.boot.n.fixedB.B100000.uiv.p2.rcr-vs-generic-ppi-source.p-holm\endcsname{$+$13.0}
\expandafter\def\csname val@rev.pol.boot.n.fixedB.B100000.uiv.p2.rcr-vs-generic-ppi-source.p-raw\endcsname{0.043}
\expandafter\def\csname cnt@rev.pol.boot.n.fixedB.B100000.uiv.p2.rcr-vs-generic-ppi-source.p-raw\endcsname{0}
\expandafter\def\csname pct@rev.pol.boot.n.fixedB.B100000.uiv.p2.rcr-vs-generic-ppi-source.p-raw\endcsname{4.3}
\expandafter\def\csname ppt@rev.pol.boot.n.fixedB.B100000.uiv.p2.rcr-vs-generic-ppi-source.p-raw\endcsname{$+$4.3}
\expandafter\def\csname val@rev.pol.boot.n.fixedB.B100000.uiv.p2.rcr-vs-generic-ppi-source.reject\endcsname{0}
\expandafter\def\csname cnt@rev.pol.boot.n.fixedB.B100000.uiv.p2.rcr-vs-generic-ppi-source.reject\endcsname{0}
\expandafter\def\csname pct@rev.pol.boot.n.fixedB.B100000.uiv.p2.rcr-vs-generic-ppi-source.reject\endcsname{0.0}
\expandafter\def\csname ppt@rev.pol.boot.n.fixedB.B100000.uiv.p2.rcr-vs-generic-ppi-source.reject\endcsname{0.0}
\expandafter\def\csname val@rev.pol.boot.n.fixedB.B100000.uiv.p2.rcr-vs-replay-ppi-uniform.p-holm\endcsname{0.017}
\expandafter\def\csname cnt@rev.pol.boot.n.fixedB.B100000.uiv.p2.rcr-vs-replay-ppi-uniform.p-holm\endcsname{0}
\expandafter\def\csname pct@rev.pol.boot.n.fixedB.B100000.uiv.p2.rcr-vs-replay-ppi-uniform.p-holm\endcsname{1.7}
\expandafter\def\csname ppt@rev.pol.boot.n.fixedB.B100000.uiv.p2.rcr-vs-replay-ppi-uniform.p-holm\endcsname{$+$1.7}
\expandafter\def\csname val@rev.pol.boot.n.fixedB.B100000.uiv.p2.rcr-vs-replay-ppi-uniform.p-raw\endcsname{0.003}
\expandafter\def\csname cnt@rev.pol.boot.n.fixedB.B100000.uiv.p2.rcr-vs-replay-ppi-uniform.p-raw\endcsname{0}
\expandafter\def\csname pct@rev.pol.boot.n.fixedB.B100000.uiv.p2.rcr-vs-replay-ppi-uniform.p-raw\endcsname{0.3}
\expandafter\def\csname ppt@rev.pol.boot.n.fixedB.B100000.uiv.p2.rcr-vs-replay-ppi-uniform.p-raw\endcsname{$+$0.3}
\expandafter\def\csname val@rev.pol.boot.n.fixedB.B100000.uiv.p2.rcr-vs-replay-ppi-uniform.reject\endcsname{1}
\expandafter\def\csname cnt@rev.pol.boot.n.fixedB.B100000.uiv.p2.rcr-vs-replay-ppi-uniform.reject\endcsname{1}
\expandafter\def\csname pct@rev.pol.boot.n.fixedB.B100000.uiv.p2.rcr-vs-replay-ppi-uniform.reject\endcsname{100.0}
\expandafter\def\csname ppt@rev.pol.boot.n.fixedB.B100000.uiv.p2.rcr-vs-replay-ppi-uniform.reject\endcsname{$+$100.0}
\expandafter\def\csname val@rev.pol.boot.n.fixedB.B4000.n-rejected\endcsname{5}
\expandafter\def\csname cnt@rev.pol.boot.n.fixedB.B4000.n-rejected\endcsname{5}
\expandafter\def\csname pct@rev.pol.boot.n.fixedB.B4000.n-rejected\endcsname{500.0}
\expandafter\def\csname ppt@rev.pol.boot.n.fixedB.B4000.n-rejected\endcsname{$+$500.0}
\expandafter\def\csname val@rev.pol.boot.n.fixedB.B4000.sspro.p1.rcr-vs-generic-ppi-source.p-holm\endcsname{0.035}
\expandafter\def\csname cnt@rev.pol.boot.n.fixedB.B4000.sspro.p1.rcr-vs-generic-ppi-source.p-holm\endcsname{0}
\expandafter\def\csname pct@rev.pol.boot.n.fixedB.B4000.sspro.p1.rcr-vs-generic-ppi-source.p-holm\endcsname{3.5}
\expandafter\def\csname ppt@rev.pol.boot.n.fixedB.B4000.sspro.p1.rcr-vs-generic-ppi-source.p-holm\endcsname{$+$3.5}
\expandafter\def\csname val@rev.pol.boot.n.fixedB.B4000.sspro.p1.rcr-vs-generic-ppi-source.p-raw\endcsname{0.007}
\expandafter\def\csname cnt@rev.pol.boot.n.fixedB.B4000.sspro.p1.rcr-vs-generic-ppi-source.p-raw\endcsname{0}
\expandafter\def\csname pct@rev.pol.boot.n.fixedB.B4000.sspro.p1.rcr-vs-generic-ppi-source.p-raw\endcsname{0.7}
\expandafter\def\csname ppt@rev.pol.boot.n.fixedB.B4000.sspro.p1.rcr-vs-generic-ppi-source.p-raw\endcsname{$+$0.7}
\expandafter\def\csname val@rev.pol.boot.n.fixedB.B4000.sspro.p1.rcr-vs-generic-ppi-source.reject\endcsname{1}
\expandafter\def\csname cnt@rev.pol.boot.n.fixedB.B4000.sspro.p1.rcr-vs-generic-ppi-source.reject\endcsname{1}
\expandafter\def\csname pct@rev.pol.boot.n.fixedB.B4000.sspro.p1.rcr-vs-generic-ppi-source.reject\endcsname{100.0}
\expandafter\def\csname ppt@rev.pol.boot.n.fixedB.B4000.sspro.p1.rcr-vs-generic-ppi-source.reject\endcsname{$+$100.0}
\expandafter\def\csname val@rev.pol.boot.n.fixedB.B4000.sspro.p1.rcr-vs-replay-ppi-uniform.p-holm\endcsname{0.038}
\expandafter\def\csname cnt@rev.pol.boot.n.fixedB.B4000.sspro.p1.rcr-vs-replay-ppi-uniform.p-holm\endcsname{0}
\expandafter\def\csname pct@rev.pol.boot.n.fixedB.B4000.sspro.p1.rcr-vs-replay-ppi-uniform.p-holm\endcsname{3.8}
\expandafter\def\csname ppt@rev.pol.boot.n.fixedB.B4000.sspro.p1.rcr-vs-replay-ppi-uniform.p-holm\endcsname{$+$3.8}
\expandafter\def\csname val@rev.pol.boot.n.fixedB.B4000.sspro.p1.rcr-vs-replay-ppi-uniform.p-raw\endcsname{0.009}
\expandafter\def\csname cnt@rev.pol.boot.n.fixedB.B4000.sspro.p1.rcr-vs-replay-ppi-uniform.p-raw\endcsname{0}
\expandafter\def\csname pct@rev.pol.boot.n.fixedB.B4000.sspro.p1.rcr-vs-replay-ppi-uniform.p-raw\endcsname{0.9}
\expandafter\def\csname ppt@rev.pol.boot.n.fixedB.B4000.sspro.p1.rcr-vs-replay-ppi-uniform.p-raw\endcsname{$+$0.9}
\expandafter\def\csname val@rev.pol.boot.n.fixedB.B4000.sspro.p1.rcr-vs-replay-ppi-uniform.reject\endcsname{1}
\expandafter\def\csname cnt@rev.pol.boot.n.fixedB.B4000.sspro.p1.rcr-vs-replay-ppi-uniform.reject\endcsname{1}
\expandafter\def\csname pct@rev.pol.boot.n.fixedB.B4000.sspro.p1.rcr-vs-replay-ppi-uniform.reject\endcsname{100.0}
\expandafter\def\csname ppt@rev.pol.boot.n.fixedB.B4000.sspro.p1.rcr-vs-replay-ppi-uniform.reject\endcsname{$+$100.0}
\expandafter\def\csname val@rev.pol.boot.n.fixedB.B4000.sspro.p2.rcr-vs-generic-ppi-source.p-holm\endcsname{0.900}
\expandafter\def\csname cnt@rev.pol.boot.n.fixedB.B4000.sspro.p2.rcr-vs-generic-ppi-source.p-holm\endcsname{1}
\expandafter\def\csname pct@rev.pol.boot.n.fixedB.B4000.sspro.p2.rcr-vs-generic-ppi-source.p-holm\endcsname{90.0}
\expandafter\def\csname ppt@rev.pol.boot.n.fixedB.B4000.sspro.p2.rcr-vs-generic-ppi-source.p-holm\endcsname{$+$90.0}
\expandafter\def\csname val@rev.pol.boot.n.fixedB.B4000.sspro.p2.rcr-vs-generic-ppi-source.p-raw\endcsname{0.585}
\expandafter\def\csname cnt@rev.pol.boot.n.fixedB.B4000.sspro.p2.rcr-vs-generic-ppi-source.p-raw\endcsname{1}
\expandafter\def\csname pct@rev.pol.boot.n.fixedB.B4000.sspro.p2.rcr-vs-generic-ppi-source.p-raw\endcsname{58.5}
\expandafter\def\csname ppt@rev.pol.boot.n.fixedB.B4000.sspro.p2.rcr-vs-generic-ppi-source.p-raw\endcsname{$+$58.5}
\expandafter\def\csname val@rev.pol.boot.n.fixedB.B4000.sspro.p2.rcr-vs-generic-ppi-source.reject\endcsname{0}
\expandafter\def\csname cnt@rev.pol.boot.n.fixedB.B4000.sspro.p2.rcr-vs-generic-ppi-source.reject\endcsname{0}
\expandafter\def\csname pct@rev.pol.boot.n.fixedB.B4000.sspro.p2.rcr-vs-generic-ppi-source.reject\endcsname{0.0}
\expandafter\def\csname ppt@rev.pol.boot.n.fixedB.B4000.sspro.p2.rcr-vs-generic-ppi-source.reject\endcsname{0.0}
\expandafter\def\csname val@rev.pol.boot.n.fixedB.B4000.sspro.p2.rcr-vs-replay-ppi-uniform.p-holm\endcsname{0.900}
\expandafter\def\csname cnt@rev.pol.boot.n.fixedB.B4000.sspro.p2.rcr-vs-replay-ppi-uniform.p-holm\endcsname{1}
\expandafter\def\csname pct@rev.pol.boot.n.fixedB.B4000.sspro.p2.rcr-vs-replay-ppi-uniform.p-holm\endcsname{90.0}
\expandafter\def\csname ppt@rev.pol.boot.n.fixedB.B4000.sspro.p2.rcr-vs-replay-ppi-uniform.p-holm\endcsname{$+$90.0}
\expandafter\def\csname val@rev.pol.boot.n.fixedB.B4000.sspro.p2.rcr-vs-replay-ppi-uniform.p-raw\endcsname{0.450}
\expandafter\def\csname cnt@rev.pol.boot.n.fixedB.B4000.sspro.p2.rcr-vs-replay-ppi-uniform.p-raw\endcsname{0}
\expandafter\def\csname pct@rev.pol.boot.n.fixedB.B4000.sspro.p2.rcr-vs-replay-ppi-uniform.p-raw\endcsname{45.0}
\expandafter\def\csname ppt@rev.pol.boot.n.fixedB.B4000.sspro.p2.rcr-vs-replay-ppi-uniform.p-raw\endcsname{$+$45.0}
\expandafter\def\csname val@rev.pol.boot.n.fixedB.B4000.sspro.p2.rcr-vs-replay-ppi-uniform.reject\endcsname{0}
\expandafter\def\csname cnt@rev.pol.boot.n.fixedB.B4000.sspro.p2.rcr-vs-replay-ppi-uniform.reject\endcsname{0}
\expandafter\def\csname pct@rev.pol.boot.n.fixedB.B4000.sspro.p2.rcr-vs-replay-ppi-uniform.reject\endcsname{0.0}
\expandafter\def\csname ppt@rev.pol.boot.n.fixedB.B4000.sspro.p2.rcr-vs-replay-ppi-uniform.reject\endcsname{0.0}
\expandafter\def\csname val@rev.pol.boot.n.fixedB.B4000.uiv.p1.rcr-vs-generic-ppi-source.p-holm\endcsname{0.003}
\expandafter\def\csname cnt@rev.pol.boot.n.fixedB.B4000.uiv.p1.rcr-vs-generic-ppi-source.p-holm\endcsname{0}
\expandafter\def\csname pct@rev.pol.boot.n.fixedB.B4000.uiv.p1.rcr-vs-generic-ppi-source.p-holm\endcsname{0.3}
\expandafter\def\csname ppt@rev.pol.boot.n.fixedB.B4000.uiv.p1.rcr-vs-generic-ppi-source.p-holm\endcsname{$+$0.3}
\expandafter\def\csname val@rev.pol.boot.n.fixedB.B4000.uiv.p1.rcr-vs-generic-ppi-source.p-raw\endcsname{0.000}
\expandafter\def\csname cnt@rev.pol.boot.n.fixedB.B4000.uiv.p1.rcr-vs-generic-ppi-source.p-raw\endcsname{0}
\expandafter\def\csname pct@rev.pol.boot.n.fixedB.B4000.uiv.p1.rcr-vs-generic-ppi-source.p-raw\endcsname{0.0}
\expandafter\def\csname ppt@rev.pol.boot.n.fixedB.B4000.uiv.p1.rcr-vs-generic-ppi-source.p-raw\endcsname{0.0}
\expandafter\def\csname val@rev.pol.boot.n.fixedB.B4000.uiv.p1.rcr-vs-generic-ppi-source.reject\endcsname{1}
\expandafter\def\csname cnt@rev.pol.boot.n.fixedB.B4000.uiv.p1.rcr-vs-generic-ppi-source.reject\endcsname{1}
\expandafter\def\csname pct@rev.pol.boot.n.fixedB.B4000.uiv.p1.rcr-vs-generic-ppi-source.reject\endcsname{100.0}
\expandafter\def\csname ppt@rev.pol.boot.n.fixedB.B4000.uiv.p1.rcr-vs-generic-ppi-source.reject\endcsname{$+$100.0}
\expandafter\def\csname val@rev.pol.boot.n.fixedB.B4000.uiv.p1.rcr-vs-replay-ppi-uniform.p-holm\endcsname{0.002}
\expandafter\def\csname cnt@rev.pol.boot.n.fixedB.B4000.uiv.p1.rcr-vs-replay-ppi-uniform.p-holm\endcsname{0}
\expandafter\def\csname pct@rev.pol.boot.n.fixedB.B4000.uiv.p1.rcr-vs-replay-ppi-uniform.p-holm\endcsname{0.2}
\expandafter\def\csname ppt@rev.pol.boot.n.fixedB.B4000.uiv.p1.rcr-vs-replay-ppi-uniform.p-holm\endcsname{$+$0.2}
\expandafter\def\csname val@rev.pol.boot.n.fixedB.B4000.uiv.p1.rcr-vs-replay-ppi-uniform.p-raw\endcsname{0.000}
\expandafter\def\csname cnt@rev.pol.boot.n.fixedB.B4000.uiv.p1.rcr-vs-replay-ppi-uniform.p-raw\endcsname{0}
\expandafter\def\csname pct@rev.pol.boot.n.fixedB.B4000.uiv.p1.rcr-vs-replay-ppi-uniform.p-raw\endcsname{0.0}
\expandafter\def\csname ppt@rev.pol.boot.n.fixedB.B4000.uiv.p1.rcr-vs-replay-ppi-uniform.p-raw\endcsname{0.0}
\expandafter\def\csname val@rev.pol.boot.n.fixedB.B4000.uiv.p1.rcr-vs-replay-ppi-uniform.reject\endcsname{1}
\expandafter\def\csname cnt@rev.pol.boot.n.fixedB.B4000.uiv.p1.rcr-vs-replay-ppi-uniform.reject\endcsname{1}
\expandafter\def\csname pct@rev.pol.boot.n.fixedB.B4000.uiv.p1.rcr-vs-replay-ppi-uniform.reject\endcsname{100.0}
\expandafter\def\csname ppt@rev.pol.boot.n.fixedB.B4000.uiv.p1.rcr-vs-replay-ppi-uniform.reject\endcsname{$+$100.0}
\expandafter\def\csname val@rev.pol.boot.n.fixedB.B4000.uiv.p2.rcr-vs-generic-ppi-source.p-holm\endcsname{0.142}
\expandafter\def\csname cnt@rev.pol.boot.n.fixedB.B4000.uiv.p2.rcr-vs-generic-ppi-source.p-holm\endcsname{0}
\expandafter\def\csname pct@rev.pol.boot.n.fixedB.B4000.uiv.p2.rcr-vs-generic-ppi-source.p-holm\endcsname{14.2}
\expandafter\def\csname ppt@rev.pol.boot.n.fixedB.B4000.uiv.p2.rcr-vs-generic-ppi-source.p-holm\endcsname{$+$14.2}
\expandafter\def\csname val@rev.pol.boot.n.fixedB.B4000.uiv.p2.rcr-vs-generic-ppi-source.p-raw\endcsname{0.047}
\expandafter\def\csname cnt@rev.pol.boot.n.fixedB.B4000.uiv.p2.rcr-vs-generic-ppi-source.p-raw\endcsname{0}
\expandafter\def\csname pct@rev.pol.boot.n.fixedB.B4000.uiv.p2.rcr-vs-generic-ppi-source.p-raw\endcsname{4.7}
\expandafter\def\csname ppt@rev.pol.boot.n.fixedB.B4000.uiv.p2.rcr-vs-generic-ppi-source.p-raw\endcsname{$+$4.7}
\expandafter\def\csname val@rev.pol.boot.n.fixedB.B4000.uiv.p2.rcr-vs-generic-ppi-source.reject\endcsname{0}
\expandafter\def\csname cnt@rev.pol.boot.n.fixedB.B4000.uiv.p2.rcr-vs-generic-ppi-source.reject\endcsname{0}
\expandafter\def\csname pct@rev.pol.boot.n.fixedB.B4000.uiv.p2.rcr-vs-generic-ppi-source.reject\endcsname{0.0}
\expandafter\def\csname ppt@rev.pol.boot.n.fixedB.B4000.uiv.p2.rcr-vs-generic-ppi-source.reject\endcsname{0.0}
\expandafter\def\csname val@rev.pol.boot.n.fixedB.B4000.uiv.p2.rcr-vs-replay-ppi-uniform.p-holm\endcsname{0.016}
\expandafter\def\csname cnt@rev.pol.boot.n.fixedB.B4000.uiv.p2.rcr-vs-replay-ppi-uniform.p-holm\endcsname{0}
\expandafter\def\csname pct@rev.pol.boot.n.fixedB.B4000.uiv.p2.rcr-vs-replay-ppi-uniform.p-holm\endcsname{1.6}
\expandafter\def\csname ppt@rev.pol.boot.n.fixedB.B4000.uiv.p2.rcr-vs-replay-ppi-uniform.p-holm\endcsname{$+$1.6}
\expandafter\def\csname val@rev.pol.boot.n.fixedB.B4000.uiv.p2.rcr-vs-replay-ppi-uniform.p-raw\endcsname{0.003}
\expandafter\def\csname cnt@rev.pol.boot.n.fixedB.B4000.uiv.p2.rcr-vs-replay-ppi-uniform.p-raw\endcsname{0}
\expandafter\def\csname pct@rev.pol.boot.n.fixedB.B4000.uiv.p2.rcr-vs-replay-ppi-uniform.p-raw\endcsname{0.3}
\expandafter\def\csname ppt@rev.pol.boot.n.fixedB.B4000.uiv.p2.rcr-vs-replay-ppi-uniform.p-raw\endcsname{$+$0.3}
\expandafter\def\csname val@rev.pol.boot.n.fixedB.B4000.uiv.p2.rcr-vs-replay-ppi-uniform.reject\endcsname{1}
\expandafter\def\csname cnt@rev.pol.boot.n.fixedB.B4000.uiv.p2.rcr-vs-replay-ppi-uniform.reject\endcsname{1}
\expandafter\def\csname pct@rev.pol.boot.n.fixedB.B4000.uiv.p2.rcr-vs-replay-ppi-uniform.reject\endcsname{100.0}
\expandafter\def\csname ppt@rev.pol.boot.n.fixedB.B4000.uiv.p2.rcr-vs-replay-ppi-uniform.reject\endcsname{$+$100.0}
\expandafter\def\csname val@rev.pol.cat.max-abs-delta\endcsname{0.014}
\expandafter\def\csname cnt@rev.pol.cat.max-abs-delta\endcsname{0}
\expandafter\def\csname pct@rev.pol.cat.max-abs-delta\endcsname{1.4}
\expandafter\def\csname ppt@rev.pol.cat.max-abs-delta\endcsname{$+$1.4}
\expandafter\def\csname val@rev.pol.cat.sspro.p1.best-single@0.1.delta\endcsname{-0.006}
\expandafter\def\csname cnt@rev.pol.cat.sspro.p1.best-single@0.1.delta\endcsname{0}
\expandafter\def\csname pct@rev.pol.cat.sspro.p1.best-single@0.1.delta\endcsname{$-$0.6}
\expandafter\def\csname ppt@rev.pol.cat.sspro.p1.best-single@0.1.delta\endcsname{$-$0.6}
\expandafter\def\csname val@rev.pol.cat.sspro.p1.best-single@0.1.mean\endcsname{0.830}
\expandafter\def\csname cnt@rev.pol.cat.sspro.p1.best-single@0.1.mean\endcsname{1}
\expandafter\def\csname pct@rev.pol.cat.sspro.p1.best-single@0.1.mean\endcsname{83.0}
\expandafter\def\csname ppt@rev.pol.cat.sspro.p1.best-single@0.1.mean\endcsname{$+$83.0}
\expandafter\def\csname val@rev.pol.cat.sspro.p1.best-single@0.1.mean20\endcsname{0.836}
\expandafter\def\csname cnt@rev.pol.cat.sspro.p1.best-single@0.1.mean20\endcsname{1}
\expandafter\def\csname pct@rev.pol.cat.sspro.p1.best-single@0.1.mean20\endcsname{83.6}
\expandafter\def\csname ppt@rev.pol.cat.sspro.p1.best-single@0.1.mean20\endcsname{$+$83.6}
\expandafter\def\csname val@rev.pol.cat.sspro.p1.best-single@0.1.sd\endcsname{0.042}
\expandafter\def\csname cnt@rev.pol.cat.sspro.p1.best-single@0.1.sd\endcsname{0}
\expandafter\def\csname pct@rev.pol.cat.sspro.p1.best-single@0.1.sd\endcsname{4.2}
\expandafter\def\csname ppt@rev.pol.cat.sspro.p1.best-single@0.1.sd\endcsname{$+$4.2}
\expandafter\def\csname val@rev.pol.cat.sspro.p1.best-single@0.1.sd20\endcsname{0.034}
\expandafter\def\csname cnt@rev.pol.cat.sspro.p1.best-single@0.1.sd20\endcsname{0}
\expandafter\def\csname pct@rev.pol.cat.sspro.p1.best-single@0.1.sd20\endcsname{3.4}
\expandafter\def\csname ppt@rev.pol.cat.sspro.p1.best-single@0.1.sd20\endcsname{$+$3.4}
\expandafter\def\csname val@rev.pol.cat.sspro.p1.best-single@0.25.delta\endcsname{-0.014}
\expandafter\def\csname cnt@rev.pol.cat.sspro.p1.best-single@0.25.delta\endcsname{0}
\expandafter\def\csname pct@rev.pol.cat.sspro.p1.best-single@0.25.delta\endcsname{$-$1.4}
\expandafter\def\csname ppt@rev.pol.cat.sspro.p1.best-single@0.25.delta\endcsname{$-$1.4}
\expandafter\def\csname val@rev.pol.cat.sspro.p1.best-single@0.25.mean\endcsname{0.839}
\expandafter\def\csname cnt@rev.pol.cat.sspro.p1.best-single@0.25.mean\endcsname{1}
\expandafter\def\csname pct@rev.pol.cat.sspro.p1.best-single@0.25.mean\endcsname{83.9}
\expandafter\def\csname ppt@rev.pol.cat.sspro.p1.best-single@0.25.mean\endcsname{$+$83.9}
\expandafter\def\csname val@rev.pol.cat.sspro.p1.best-single@0.25.mean20\endcsname{0.853}
\expandafter\def\csname cnt@rev.pol.cat.sspro.p1.best-single@0.25.mean20\endcsname{1}
\expandafter\def\csname pct@rev.pol.cat.sspro.p1.best-single@0.25.mean20\endcsname{85.3}
\expandafter\def\csname ppt@rev.pol.cat.sspro.p1.best-single@0.25.mean20\endcsname{$+$85.3}
\expandafter\def\csname val@rev.pol.cat.sspro.p1.best-single@0.25.sd\endcsname{0.034}
\expandafter\def\csname cnt@rev.pol.cat.sspro.p1.best-single@0.25.sd\endcsname{0}
\expandafter\def\csname pct@rev.pol.cat.sspro.p1.best-single@0.25.sd\endcsname{3.4}
\expandafter\def\csname ppt@rev.pol.cat.sspro.p1.best-single@0.25.sd\endcsname{$+$3.4}
\expandafter\def\csname val@rev.pol.cat.sspro.p1.best-single@0.25.sd20\endcsname{0.016}
\expandafter\def\csname cnt@rev.pol.cat.sspro.p1.best-single@0.25.sd20\endcsname{0}
\expandafter\def\csname pct@rev.pol.cat.sspro.p1.best-single@0.25.sd20\endcsname{1.6}
\expandafter\def\csname ppt@rev.pol.cat.sspro.p1.best-single@0.25.sd20\endcsname{$+$1.6}
\expandafter\def\csname val@rev.pol.cat.sspro.p1.best-single@0.5.delta\endcsname{-0.004}
\expandafter\def\csname cnt@rev.pol.cat.sspro.p1.best-single@0.5.delta\endcsname{0}
\expandafter\def\csname pct@rev.pol.cat.sspro.p1.best-single@0.5.delta\endcsname{$-$0.4}
\expandafter\def\csname ppt@rev.pol.cat.sspro.p1.best-single@0.5.delta\endcsname{$-$0.4}
\expandafter\def\csname val@rev.pol.cat.sspro.p1.best-single@0.5.mean\endcsname{0.847}
\expandafter\def\csname cnt@rev.pol.cat.sspro.p1.best-single@0.5.mean\endcsname{1}
\expandafter\def\csname pct@rev.pol.cat.sspro.p1.best-single@0.5.mean\endcsname{84.7}
\expandafter\def\csname ppt@rev.pol.cat.sspro.p1.best-single@0.5.mean\endcsname{$+$84.7}
\expandafter\def\csname val@rev.pol.cat.sspro.p1.best-single@0.5.mean20\endcsname{0.851}
\expandafter\def\csname cnt@rev.pol.cat.sspro.p1.best-single@0.5.mean20\endcsname{1}
\expandafter\def\csname pct@rev.pol.cat.sspro.p1.best-single@0.5.mean20\endcsname{85.1}
\expandafter\def\csname ppt@rev.pol.cat.sspro.p1.best-single@0.5.mean20\endcsname{$+$85.1}
\expandafter\def\csname val@rev.pol.cat.sspro.p1.best-single@0.5.sd\endcsname{0.012}
\expandafter\def\csname cnt@rev.pol.cat.sspro.p1.best-single@0.5.sd\endcsname{0}
\expandafter\def\csname pct@rev.pol.cat.sspro.p1.best-single@0.5.sd\endcsname{1.2}
\expandafter\def\csname ppt@rev.pol.cat.sspro.p1.best-single@0.5.sd\endcsname{$+$1.2}
\expandafter\def\csname val@rev.pol.cat.sspro.p1.best-single@0.5.sd20\endcsname{0.015}
\expandafter\def\csname cnt@rev.pol.cat.sspro.p1.best-single@0.5.sd20\endcsname{0}
\expandafter\def\csname pct@rev.pol.cat.sspro.p1.best-single@0.5.sd20\endcsname{1.5}
\expandafter\def\csname ppt@rev.pol.cat.sspro.p1.best-single@0.5.sd20\endcsname{$+$1.5}
\expandafter\def\csname val@rev.pol.cat.sspro.p1.best-single@1.0.delta\endcsname{0.000}
\expandafter\def\csname cnt@rev.pol.cat.sspro.p1.best-single@1.0.delta\endcsname{0}
\expandafter\def\csname pct@rev.pol.cat.sspro.p1.best-single@1.0.delta\endcsname{0.0}
\expandafter\def\csname ppt@rev.pol.cat.sspro.p1.best-single@1.0.delta\endcsname{0.0}
\expandafter\def\csname val@rev.pol.cat.sspro.p1.best-single@1.0.mean\endcsname{0.844}
\expandafter\def\csname cnt@rev.pol.cat.sspro.p1.best-single@1.0.mean\endcsname{1}
\expandafter\def\csname pct@rev.pol.cat.sspro.p1.best-single@1.0.mean\endcsname{84.4}
\expandafter\def\csname ppt@rev.pol.cat.sspro.p1.best-single@1.0.mean\endcsname{$+$84.4}
\expandafter\def\csname val@rev.pol.cat.sspro.p1.best-single@1.0.mean20\endcsname{0.844}
\expandafter\def\csname cnt@rev.pol.cat.sspro.p1.best-single@1.0.mean20\endcsname{1}
\expandafter\def\csname pct@rev.pol.cat.sspro.p1.best-single@1.0.mean20\endcsname{84.4}
\expandafter\def\csname ppt@rev.pol.cat.sspro.p1.best-single@1.0.mean20\endcsname{$+$84.4}
\expandafter\def\csname val@rev.pol.cat.sspro.p1.best-single@1.0.sd\endcsname{0.000}
\expandafter\def\csname cnt@rev.pol.cat.sspro.p1.best-single@1.0.sd\endcsname{0}
\expandafter\def\csname pct@rev.pol.cat.sspro.p1.best-single@1.0.sd\endcsname{0.0}
\expandafter\def\csname ppt@rev.pol.cat.sspro.p1.best-single@1.0.sd\endcsname{0.0}
\expandafter\def\csname val@rev.pol.cat.sspro.p1.best-single@1.0.sd20\endcsname{0.000}
\expandafter\def\csname cnt@rev.pol.cat.sspro.p1.best-single@1.0.sd20\endcsname{0}
\expandafter\def\csname pct@rev.pol.cat.sspro.p1.best-single@1.0.sd20\endcsname{0.0}
\expandafter\def\csname ppt@rev.pol.cat.sspro.p1.best-single@1.0.sd20\endcsname{0.0}
\expandafter\def\csname val@rev.pol.cat.sspro.p1.generic-ppi-source@0.1.delta\endcsname{-0.004}
\expandafter\def\csname cnt@rev.pol.cat.sspro.p1.generic-ppi-source@0.1.delta\endcsname{0}
\expandafter\def\csname pct@rev.pol.cat.sspro.p1.generic-ppi-source@0.1.delta\endcsname{$-$0.4}
\expandafter\def\csname ppt@rev.pol.cat.sspro.p1.generic-ppi-source@0.1.delta\endcsname{$-$0.4}
\expandafter\def\csname val@rev.pol.cat.sspro.p1.generic-ppi-source@0.1.mean\endcsname{0.845}
\expandafter\def\csname cnt@rev.pol.cat.sspro.p1.generic-ppi-source@0.1.mean\endcsname{1}
\expandafter\def\csname pct@rev.pol.cat.sspro.p1.generic-ppi-source@0.1.mean\endcsname{84.5}
\expandafter\def\csname ppt@rev.pol.cat.sspro.p1.generic-ppi-source@0.1.mean\endcsname{$+$84.5}
\expandafter\def\csname val@rev.pol.cat.sspro.p1.generic-ppi-source@0.1.mean20\endcsname{0.849}
\expandafter\def\csname cnt@rev.pol.cat.sspro.p1.generic-ppi-source@0.1.mean20\endcsname{1}
\expandafter\def\csname pct@rev.pol.cat.sspro.p1.generic-ppi-source@0.1.mean20\endcsname{84.9}
\expandafter\def\csname ppt@rev.pol.cat.sspro.p1.generic-ppi-source@0.1.mean20\endcsname{$+$84.9}
\expandafter\def\csname val@rev.pol.cat.sspro.p1.generic-ppi-source@0.1.sd\endcsname{0.024}
\expandafter\def\csname cnt@rev.pol.cat.sspro.p1.generic-ppi-source@0.1.sd\endcsname{0}
\expandafter\def\csname pct@rev.pol.cat.sspro.p1.generic-ppi-source@0.1.sd\endcsname{2.4}
\expandafter\def\csname ppt@rev.pol.cat.sspro.p1.generic-ppi-source@0.1.sd\endcsname{$+$2.4}
\expandafter\def\csname val@rev.pol.cat.sspro.p1.generic-ppi-source@0.1.sd20\endcsname{0.017}
\expandafter\def\csname cnt@rev.pol.cat.sspro.p1.generic-ppi-source@0.1.sd20\endcsname{0}
\expandafter\def\csname pct@rev.pol.cat.sspro.p1.generic-ppi-source@0.1.sd20\endcsname{1.7}
\expandafter\def\csname ppt@rev.pol.cat.sspro.p1.generic-ppi-source@0.1.sd20\endcsname{$+$1.7}
\expandafter\def\csname val@rev.pol.cat.sspro.p1.generic-ppi-source@0.25.delta\endcsname{-0.004}
\expandafter\def\csname cnt@rev.pol.cat.sspro.p1.generic-ppi-source@0.25.delta\endcsname{0}
\expandafter\def\csname pct@rev.pol.cat.sspro.p1.generic-ppi-source@0.25.delta\endcsname{$-$0.4}
\expandafter\def\csname ppt@rev.pol.cat.sspro.p1.generic-ppi-source@0.25.delta\endcsname{$-$0.4}
\expandafter\def\csname val@rev.pol.cat.sspro.p1.generic-ppi-source@0.25.mean\endcsname{0.847}
\expandafter\def\csname cnt@rev.pol.cat.sspro.p1.generic-ppi-source@0.25.mean\endcsname{1}
\expandafter\def\csname pct@rev.pol.cat.sspro.p1.generic-ppi-source@0.25.mean\endcsname{84.7}
\expandafter\def\csname ppt@rev.pol.cat.sspro.p1.generic-ppi-source@0.25.mean\endcsname{$+$84.7}
\expandafter\def\csname val@rev.pol.cat.sspro.p1.generic-ppi-source@0.25.mean20\endcsname{0.852}
\expandafter\def\csname cnt@rev.pol.cat.sspro.p1.generic-ppi-source@0.25.mean20\endcsname{1}
\expandafter\def\csname pct@rev.pol.cat.sspro.p1.generic-ppi-source@0.25.mean20\endcsname{85.2}
\expandafter\def\csname ppt@rev.pol.cat.sspro.p1.generic-ppi-source@0.25.mean20\endcsname{$+$85.2}
\expandafter\def\csname val@rev.pol.cat.sspro.p1.generic-ppi-source@0.25.sd\endcsname{0.020}
\expandafter\def\csname cnt@rev.pol.cat.sspro.p1.generic-ppi-source@0.25.sd\endcsname{0}
\expandafter\def\csname pct@rev.pol.cat.sspro.p1.generic-ppi-source@0.25.sd\endcsname{2.0}
\expandafter\def\csname ppt@rev.pol.cat.sspro.p1.generic-ppi-source@0.25.sd\endcsname{$+$2.0}
\expandafter\def\csname val@rev.pol.cat.sspro.p1.generic-ppi-source@0.25.sd20\endcsname{0.014}
\expandafter\def\csname cnt@rev.pol.cat.sspro.p1.generic-ppi-source@0.25.sd20\endcsname{0}
\expandafter\def\csname pct@rev.pol.cat.sspro.p1.generic-ppi-source@0.25.sd20\endcsname{1.4}
\expandafter\def\csname ppt@rev.pol.cat.sspro.p1.generic-ppi-source@0.25.sd20\endcsname{$+$1.4}
\expandafter\def\csname val@rev.pol.cat.sspro.p1.generic-ppi-source@0.5.delta\endcsname{-0.001}
\expandafter\def\csname cnt@rev.pol.cat.sspro.p1.generic-ppi-source@0.5.delta\endcsname{0}
\expandafter\def\csname pct@rev.pol.cat.sspro.p1.generic-ppi-source@0.5.delta\endcsname{$-$0.1}
\expandafter\def\csname ppt@rev.pol.cat.sspro.p1.generic-ppi-source@0.5.delta\endcsname{$-$0.1}
\expandafter\def\csname val@rev.pol.cat.sspro.p1.generic-ppi-source@0.5.mean\endcsname{0.858}
\expandafter\def\csname cnt@rev.pol.cat.sspro.p1.generic-ppi-source@0.5.mean\endcsname{1}
\expandafter\def\csname pct@rev.pol.cat.sspro.p1.generic-ppi-source@0.5.mean\endcsname{85.8}
\expandafter\def\csname ppt@rev.pol.cat.sspro.p1.generic-ppi-source@0.5.mean\endcsname{$+$85.8}
\expandafter\def\csname val@rev.pol.cat.sspro.p1.generic-ppi-source@0.5.mean20\endcsname{0.859}
\expandafter\def\csname cnt@rev.pol.cat.sspro.p1.generic-ppi-source@0.5.mean20\endcsname{1}
\expandafter\def\csname pct@rev.pol.cat.sspro.p1.generic-ppi-source@0.5.mean20\endcsname{85.9}
\expandafter\def\csname ppt@rev.pol.cat.sspro.p1.generic-ppi-source@0.5.mean20\endcsname{$+$85.9}
\expandafter\def\csname val@rev.pol.cat.sspro.p1.generic-ppi-source@0.5.sd\endcsname{0.006}
\expandafter\def\csname cnt@rev.pol.cat.sspro.p1.generic-ppi-source@0.5.sd\endcsname{0}
\expandafter\def\csname pct@rev.pol.cat.sspro.p1.generic-ppi-source@0.5.sd\endcsname{0.6}
\expandafter\def\csname ppt@rev.pol.cat.sspro.p1.generic-ppi-source@0.5.sd\endcsname{$+$0.6}
\expandafter\def\csname val@rev.pol.cat.sspro.p1.generic-ppi-source@0.5.sd20\endcsname{0.000}
\expandafter\def\csname cnt@rev.pol.cat.sspro.p1.generic-ppi-source@0.5.sd20\endcsname{0}
\expandafter\def\csname pct@rev.pol.cat.sspro.p1.generic-ppi-source@0.5.sd20\endcsname{0.0}
\expandafter\def\csname ppt@rev.pol.cat.sspro.p1.generic-ppi-source@0.5.sd20\endcsname{0.0}
\expandafter\def\csname val@rev.pol.cat.sspro.p1.generic-ppi-source@1.0.delta\endcsname{0.000}
\expandafter\def\csname cnt@rev.pol.cat.sspro.p1.generic-ppi-source@1.0.delta\endcsname{0}
\expandafter\def\csname pct@rev.pol.cat.sspro.p1.generic-ppi-source@1.0.delta\endcsname{0.0}
\expandafter\def\csname ppt@rev.pol.cat.sspro.p1.generic-ppi-source@1.0.delta\endcsname{0.0}
\expandafter\def\csname val@rev.pol.cat.sspro.p1.generic-ppi-source@1.0.mean\endcsname{0.859}
\expandafter\def\csname cnt@rev.pol.cat.sspro.p1.generic-ppi-source@1.0.mean\endcsname{1}
\expandafter\def\csname pct@rev.pol.cat.sspro.p1.generic-ppi-source@1.0.mean\endcsname{85.9}
\expandafter\def\csname ppt@rev.pol.cat.sspro.p1.generic-ppi-source@1.0.mean\endcsname{$+$85.9}
\expandafter\def\csname val@rev.pol.cat.sspro.p1.generic-ppi-source@1.0.mean20\endcsname{0.859}
\expandafter\def\csname cnt@rev.pol.cat.sspro.p1.generic-ppi-source@1.0.mean20\endcsname{1}
\expandafter\def\csname pct@rev.pol.cat.sspro.p1.generic-ppi-source@1.0.mean20\endcsname{85.9}
\expandafter\def\csname ppt@rev.pol.cat.sspro.p1.generic-ppi-source@1.0.mean20\endcsname{$+$85.9}
\expandafter\def\csname val@rev.pol.cat.sspro.p1.generic-ppi-source@1.0.sd\endcsname{0.000}
\expandafter\def\csname cnt@rev.pol.cat.sspro.p1.generic-ppi-source@1.0.sd\endcsname{0}
\expandafter\def\csname pct@rev.pol.cat.sspro.p1.generic-ppi-source@1.0.sd\endcsname{0.0}
\expandafter\def\csname ppt@rev.pol.cat.sspro.p1.generic-ppi-source@1.0.sd\endcsname{0.0}
\expandafter\def\csname val@rev.pol.cat.sspro.p1.generic-ppi-source@1.0.sd20\endcsname{0.000}
\expandafter\def\csname cnt@rev.pol.cat.sspro.p1.generic-ppi-source@1.0.sd20\endcsname{0}
\expandafter\def\csname pct@rev.pol.cat.sspro.p1.generic-ppi-source@1.0.sd20\endcsname{0.0}
\expandafter\def\csname ppt@rev.pol.cat.sspro.p1.generic-ppi-source@1.0.sd20\endcsname{0.0}
\expandafter\def\csname val@rev.pol.cat.sspro.p1.rcr-stratified@0.1.delta\endcsname{0.000}
\expandafter\def\csname cnt@rev.pol.cat.sspro.p1.rcr-stratified@0.1.delta\endcsname{0}
\expandafter\def\csname pct@rev.pol.cat.sspro.p1.rcr-stratified@0.1.delta\endcsname{0.0}
\expandafter\def\csname ppt@rev.pol.cat.sspro.p1.rcr-stratified@0.1.delta\endcsname{0.0}
\expandafter\def\csname val@rev.pol.cat.sspro.p1.rcr-stratified@0.1.mean\endcsname{0.740}
\expandafter\def\csname cnt@rev.pol.cat.sspro.p1.rcr-stratified@0.1.mean\endcsname{1}
\expandafter\def\csname pct@rev.pol.cat.sspro.p1.rcr-stratified@0.1.mean\endcsname{74.0}
\expandafter\def\csname ppt@rev.pol.cat.sspro.p1.rcr-stratified@0.1.mean\endcsname{$+$74.0}
\expandafter\def\csname val@rev.pol.cat.sspro.p1.rcr-stratified@0.1.mean20\endcsname{0.740}
\expandafter\def\csname cnt@rev.pol.cat.sspro.p1.rcr-stratified@0.1.mean20\endcsname{1}
\expandafter\def\csname pct@rev.pol.cat.sspro.p1.rcr-stratified@0.1.mean20\endcsname{74.0}
\expandafter\def\csname ppt@rev.pol.cat.sspro.p1.rcr-stratified@0.1.mean20\endcsname{$+$74.0}
\expandafter\def\csname val@rev.pol.cat.sspro.p1.rcr-stratified@0.1.sd\endcsname{0.000}
\expandafter\def\csname cnt@rev.pol.cat.sspro.p1.rcr-stratified@0.1.sd\endcsname{0}
\expandafter\def\csname pct@rev.pol.cat.sspro.p1.rcr-stratified@0.1.sd\endcsname{0.0}
\expandafter\def\csname ppt@rev.pol.cat.sspro.p1.rcr-stratified@0.1.sd\endcsname{0.0}
\expandafter\def\csname val@rev.pol.cat.sspro.p1.rcr-stratified@0.1.sd20\endcsname{0.000}
\expandafter\def\csname cnt@rev.pol.cat.sspro.p1.rcr-stratified@0.1.sd20\endcsname{0}
\expandafter\def\csname pct@rev.pol.cat.sspro.p1.rcr-stratified@0.1.sd20\endcsname{0.0}
\expandafter\def\csname ppt@rev.pol.cat.sspro.p1.rcr-stratified@0.1.sd20\endcsname{0.0}
\expandafter\def\csname val@rev.pol.cat.sspro.p1.rcr-stratified@0.25.delta\endcsname{-0.002}
\expandafter\def\csname cnt@rev.pol.cat.sspro.p1.rcr-stratified@0.25.delta\endcsname{0}
\expandafter\def\csname pct@rev.pol.cat.sspro.p1.rcr-stratified@0.25.delta\endcsname{$-$0.2}
\expandafter\def\csname ppt@rev.pol.cat.sspro.p1.rcr-stratified@0.25.delta\endcsname{$-$0.2}
\expandafter\def\csname val@rev.pol.cat.sspro.p1.rcr-stratified@0.25.mean\endcsname{0.856}
\expandafter\def\csname cnt@rev.pol.cat.sspro.p1.rcr-stratified@0.25.mean\endcsname{1}
\expandafter\def\csname pct@rev.pol.cat.sspro.p1.rcr-stratified@0.25.mean\endcsname{85.6}
\expandafter\def\csname ppt@rev.pol.cat.sspro.p1.rcr-stratified@0.25.mean\endcsname{$+$85.6}
\expandafter\def\csname val@rev.pol.cat.sspro.p1.rcr-stratified@0.25.mean20\endcsname{0.858}
\expandafter\def\csname cnt@rev.pol.cat.sspro.p1.rcr-stratified@0.25.mean20\endcsname{1}
\expandafter\def\csname pct@rev.pol.cat.sspro.p1.rcr-stratified@0.25.mean20\endcsname{85.8}
\expandafter\def\csname ppt@rev.pol.cat.sspro.p1.rcr-stratified@0.25.mean20\endcsname{$+$85.8}
\expandafter\def\csname val@rev.pol.cat.sspro.p1.rcr-stratified@0.25.sd\endcsname{0.011}
\expandafter\def\csname cnt@rev.pol.cat.sspro.p1.rcr-stratified@0.25.sd\endcsname{0}
\expandafter\def\csname pct@rev.pol.cat.sspro.p1.rcr-stratified@0.25.sd\endcsname{1.1}
\expandafter\def\csname ppt@rev.pol.cat.sspro.p1.rcr-stratified@0.25.sd\endcsname{$+$1.1}
\expandafter\def\csname val@rev.pol.cat.sspro.p1.rcr-stratified@0.25.sd20\endcsname{0.008}
\expandafter\def\csname cnt@rev.pol.cat.sspro.p1.rcr-stratified@0.25.sd20\endcsname{0}
\expandafter\def\csname pct@rev.pol.cat.sspro.p1.rcr-stratified@0.25.sd20\endcsname{0.8}
\expandafter\def\csname ppt@rev.pol.cat.sspro.p1.rcr-stratified@0.25.sd20\endcsname{$+$0.8}
\expandafter\def\csname val@rev.pol.cat.sspro.p1.rcr-stratified@0.5.delta\endcsname{-0.000}
\expandafter\def\csname cnt@rev.pol.cat.sspro.p1.rcr-stratified@0.5.delta\endcsname{0}
\expandafter\def\csname pct@rev.pol.cat.sspro.p1.rcr-stratified@0.5.delta\endcsname{0.0}
\expandafter\def\csname ppt@rev.pol.cat.sspro.p1.rcr-stratified@0.5.delta\endcsname{0.0}
\expandafter\def\csname val@rev.pol.cat.sspro.p1.rcr-stratified@0.5.mean\endcsname{0.859}
\expandafter\def\csname cnt@rev.pol.cat.sspro.p1.rcr-stratified@0.5.mean\endcsname{1}
\expandafter\def\csname pct@rev.pol.cat.sspro.p1.rcr-stratified@0.5.mean\endcsname{85.9}
\expandafter\def\csname ppt@rev.pol.cat.sspro.p1.rcr-stratified@0.5.mean\endcsname{$+$85.9}
\expandafter\def\csname val@rev.pol.cat.sspro.p1.rcr-stratified@0.5.mean20\endcsname{0.859}
\expandafter\def\csname cnt@rev.pol.cat.sspro.p1.rcr-stratified@0.5.mean20\endcsname{1}
\expandafter\def\csname pct@rev.pol.cat.sspro.p1.rcr-stratified@0.5.mean20\endcsname{85.9}
\expandafter\def\csname ppt@rev.pol.cat.sspro.p1.rcr-stratified@0.5.mean20\endcsname{$+$85.9}
\expandafter\def\csname val@rev.pol.cat.sspro.p1.rcr-stratified@0.5.sd\endcsname{0.004}
\expandafter\def\csname cnt@rev.pol.cat.sspro.p1.rcr-stratified@0.5.sd\endcsname{0}
\expandafter\def\csname pct@rev.pol.cat.sspro.p1.rcr-stratified@0.5.sd\endcsname{0.4}
\expandafter\def\csname ppt@rev.pol.cat.sspro.p1.rcr-stratified@0.5.sd\endcsname{$+$0.4}
\expandafter\def\csname val@rev.pol.cat.sspro.p1.rcr-stratified@0.5.sd20\endcsname{0.000}
\expandafter\def\csname cnt@rev.pol.cat.sspro.p1.rcr-stratified@0.5.sd20\endcsname{0}
\expandafter\def\csname pct@rev.pol.cat.sspro.p1.rcr-stratified@0.5.sd20\endcsname{0.0}
\expandafter\def\csname ppt@rev.pol.cat.sspro.p1.rcr-stratified@0.5.sd20\endcsname{0.0}
\expandafter\def\csname val@rev.pol.cat.sspro.p1.rcr-stratified@1.0.delta\endcsname{0.000}
\expandafter\def\csname cnt@rev.pol.cat.sspro.p1.rcr-stratified@1.0.delta\endcsname{0}
\expandafter\def\csname pct@rev.pol.cat.sspro.p1.rcr-stratified@1.0.delta\endcsname{0.0}
\expandafter\def\csname ppt@rev.pol.cat.sspro.p1.rcr-stratified@1.0.delta\endcsname{0.0}
\expandafter\def\csname val@rev.pol.cat.sspro.p1.rcr-stratified@1.0.mean\endcsname{0.859}
\expandafter\def\csname cnt@rev.pol.cat.sspro.p1.rcr-stratified@1.0.mean\endcsname{1}
\expandafter\def\csname pct@rev.pol.cat.sspro.p1.rcr-stratified@1.0.mean\endcsname{85.9}
\expandafter\def\csname ppt@rev.pol.cat.sspro.p1.rcr-stratified@1.0.mean\endcsname{$+$85.9}
\expandafter\def\csname val@rev.pol.cat.sspro.p1.rcr-stratified@1.0.mean20\endcsname{0.859}
\expandafter\def\csname cnt@rev.pol.cat.sspro.p1.rcr-stratified@1.0.mean20\endcsname{1}
\expandafter\def\csname pct@rev.pol.cat.sspro.p1.rcr-stratified@1.0.mean20\endcsname{85.9}
\expandafter\def\csname ppt@rev.pol.cat.sspro.p1.rcr-stratified@1.0.mean20\endcsname{$+$85.9}
\expandafter\def\csname val@rev.pol.cat.sspro.p1.rcr-stratified@1.0.sd\endcsname{0.000}
\expandafter\def\csname cnt@rev.pol.cat.sspro.p1.rcr-stratified@1.0.sd\endcsname{0}
\expandafter\def\csname pct@rev.pol.cat.sspro.p1.rcr-stratified@1.0.sd\endcsname{0.0}
\expandafter\def\csname ppt@rev.pol.cat.sspro.p1.rcr-stratified@1.0.sd\endcsname{0.0}
\expandafter\def\csname val@rev.pol.cat.sspro.p1.rcr-stratified@1.0.sd20\endcsname{0.000}
\expandafter\def\csname cnt@rev.pol.cat.sspro.p1.rcr-stratified@1.0.sd20\endcsname{0}
\expandafter\def\csname pct@rev.pol.cat.sspro.p1.rcr-stratified@1.0.sd20\endcsname{0.0}
\expandafter\def\csname ppt@rev.pol.cat.sspro.p1.rcr-stratified@1.0.sd20\endcsname{0.0}
\expandafter\def\csname val@rev.pol.cat.sspro.p1.replay-ppi-uniform@0.1.delta\endcsname{-0.004}
\expandafter\def\csname cnt@rev.pol.cat.sspro.p1.replay-ppi-uniform@0.1.delta\endcsname{0}
\expandafter\def\csname pct@rev.pol.cat.sspro.p1.replay-ppi-uniform@0.1.delta\endcsname{$-$0.4}
\expandafter\def\csname ppt@rev.pol.cat.sspro.p1.replay-ppi-uniform@0.1.delta\endcsname{$-$0.4}
\expandafter\def\csname val@rev.pol.cat.sspro.p1.replay-ppi-uniform@0.1.mean\endcsname{0.843}
\expandafter\def\csname cnt@rev.pol.cat.sspro.p1.replay-ppi-uniform@0.1.mean\endcsname{1}
\expandafter\def\csname pct@rev.pol.cat.sspro.p1.replay-ppi-uniform@0.1.mean\endcsname{84.3}
\expandafter\def\csname ppt@rev.pol.cat.sspro.p1.replay-ppi-uniform@0.1.mean\endcsname{$+$84.3}
\expandafter\def\csname val@rev.pol.cat.sspro.p1.replay-ppi-uniform@0.1.mean20\endcsname{0.847}
\expandafter\def\csname cnt@rev.pol.cat.sspro.p1.replay-ppi-uniform@0.1.mean20\endcsname{1}
\expandafter\def\csname pct@rev.pol.cat.sspro.p1.replay-ppi-uniform@0.1.mean20\endcsname{84.7}
\expandafter\def\csname ppt@rev.pol.cat.sspro.p1.replay-ppi-uniform@0.1.mean20\endcsname{$+$84.7}
\expandafter\def\csname val@rev.pol.cat.sspro.p1.replay-ppi-uniform@0.1.sd\endcsname{0.024}
\expandafter\def\csname cnt@rev.pol.cat.sspro.p1.replay-ppi-uniform@0.1.sd\endcsname{0}
\expandafter\def\csname pct@rev.pol.cat.sspro.p1.replay-ppi-uniform@0.1.sd\endcsname{2.4}
\expandafter\def\csname ppt@rev.pol.cat.sspro.p1.replay-ppi-uniform@0.1.sd\endcsname{$+$2.4}
\expandafter\def\csname val@rev.pol.cat.sspro.p1.replay-ppi-uniform@0.1.sd20\endcsname{0.018}
\expandafter\def\csname cnt@rev.pol.cat.sspro.p1.replay-ppi-uniform@0.1.sd20\endcsname{0}
\expandafter\def\csname pct@rev.pol.cat.sspro.p1.replay-ppi-uniform@0.1.sd20\endcsname{1.8}
\expandafter\def\csname ppt@rev.pol.cat.sspro.p1.replay-ppi-uniform@0.1.sd20\endcsname{$+$1.8}
\expandafter\def\csname val@rev.pol.cat.sspro.p1.replay-ppi-uniform@0.25.delta\endcsname{0.004}
\expandafter\def\csname cnt@rev.pol.cat.sspro.p1.replay-ppi-uniform@0.25.delta\endcsname{0}
\expandafter\def\csname pct@rev.pol.cat.sspro.p1.replay-ppi-uniform@0.25.delta\endcsname{0.4}
\expandafter\def\csname ppt@rev.pol.cat.sspro.p1.replay-ppi-uniform@0.25.delta\endcsname{$+$0.4}
\expandafter\def\csname val@rev.pol.cat.sspro.p1.replay-ppi-uniform@0.25.mean\endcsname{0.852}
\expandafter\def\csname cnt@rev.pol.cat.sspro.p1.replay-ppi-uniform@0.25.mean\endcsname{1}
\expandafter\def\csname pct@rev.pol.cat.sspro.p1.replay-ppi-uniform@0.25.mean\endcsname{85.2}
\expandafter\def\csname ppt@rev.pol.cat.sspro.p1.replay-ppi-uniform@0.25.mean\endcsname{$+$85.2}
\expandafter\def\csname val@rev.pol.cat.sspro.p1.replay-ppi-uniform@0.25.mean20\endcsname{0.849}
\expandafter\def\csname cnt@rev.pol.cat.sspro.p1.replay-ppi-uniform@0.25.mean20\endcsname{1}
\expandafter\def\csname pct@rev.pol.cat.sspro.p1.replay-ppi-uniform@0.25.mean20\endcsname{84.9}
\expandafter\def\csname ppt@rev.pol.cat.sspro.p1.replay-ppi-uniform@0.25.mean20\endcsname{$+$84.9}
\expandafter\def\csname val@rev.pol.cat.sspro.p1.replay-ppi-uniform@0.25.sd\endcsname{0.015}
\expandafter\def\csname cnt@rev.pol.cat.sspro.p1.replay-ppi-uniform@0.25.sd\endcsname{0}
\expandafter\def\csname pct@rev.pol.cat.sspro.p1.replay-ppi-uniform@0.25.sd\endcsname{1.5}
\expandafter\def\csname ppt@rev.pol.cat.sspro.p1.replay-ppi-uniform@0.25.sd\endcsname{$+$1.5}
\expandafter\def\csname val@rev.pol.cat.sspro.p1.replay-ppi-uniform@0.25.sd20\endcsname{0.016}
\expandafter\def\csname cnt@rev.pol.cat.sspro.p1.replay-ppi-uniform@0.25.sd20\endcsname{0}
\expandafter\def\csname pct@rev.pol.cat.sspro.p1.replay-ppi-uniform@0.25.sd20\endcsname{1.6}
\expandafter\def\csname ppt@rev.pol.cat.sspro.p1.replay-ppi-uniform@0.25.sd20\endcsname{$+$1.6}
\expandafter\def\csname val@rev.pol.cat.sspro.p1.replay-ppi-uniform@0.5.delta\endcsname{-0.000}
\expandafter\def\csname cnt@rev.pol.cat.sspro.p1.replay-ppi-uniform@0.5.delta\endcsname{0}
\expandafter\def\csname pct@rev.pol.cat.sspro.p1.replay-ppi-uniform@0.5.delta\endcsname{0.0}
\expandafter\def\csname ppt@rev.pol.cat.sspro.p1.replay-ppi-uniform@0.5.delta\endcsname{0.0}
\expandafter\def\csname val@rev.pol.cat.sspro.p1.replay-ppi-uniform@0.5.mean\endcsname{0.859}
\expandafter\def\csname cnt@rev.pol.cat.sspro.p1.replay-ppi-uniform@0.5.mean\endcsname{1}
\expandafter\def\csname pct@rev.pol.cat.sspro.p1.replay-ppi-uniform@0.5.mean\endcsname{85.9}
\expandafter\def\csname ppt@rev.pol.cat.sspro.p1.replay-ppi-uniform@0.5.mean\endcsname{$+$85.9}
\expandafter\def\csname val@rev.pol.cat.sspro.p1.replay-ppi-uniform@0.5.mean20\endcsname{0.859}
\expandafter\def\csname cnt@rev.pol.cat.sspro.p1.replay-ppi-uniform@0.5.mean20\endcsname{1}
\expandafter\def\csname pct@rev.pol.cat.sspro.p1.replay-ppi-uniform@0.5.mean20\endcsname{85.9}
\expandafter\def\csname ppt@rev.pol.cat.sspro.p1.replay-ppi-uniform@0.5.mean20\endcsname{$+$85.9}
\expandafter\def\csname val@rev.pol.cat.sspro.p1.replay-ppi-uniform@0.5.sd\endcsname{0.003}
\expandafter\def\csname cnt@rev.pol.cat.sspro.p1.replay-ppi-uniform@0.5.sd\endcsname{0}
\expandafter\def\csname pct@rev.pol.cat.sspro.p1.replay-ppi-uniform@0.5.sd\endcsname{0.3}
\expandafter\def\csname ppt@rev.pol.cat.sspro.p1.replay-ppi-uniform@0.5.sd\endcsname{$+$0.3}
\expandafter\def\csname val@rev.pol.cat.sspro.p1.replay-ppi-uniform@0.5.sd20\endcsname{0.000}
\expandafter\def\csname cnt@rev.pol.cat.sspro.p1.replay-ppi-uniform@0.5.sd20\endcsname{0}
\expandafter\def\csname pct@rev.pol.cat.sspro.p1.replay-ppi-uniform@0.5.sd20\endcsname{0.0}
\expandafter\def\csname ppt@rev.pol.cat.sspro.p1.replay-ppi-uniform@0.5.sd20\endcsname{0.0}
\expandafter\def\csname val@rev.pol.cat.sspro.p1.replay-ppi-uniform@1.0.delta\endcsname{0.000}
\expandafter\def\csname cnt@rev.pol.cat.sspro.p1.replay-ppi-uniform@1.0.delta\endcsname{0}
\expandafter\def\csname pct@rev.pol.cat.sspro.p1.replay-ppi-uniform@1.0.delta\endcsname{0.0}
\expandafter\def\csname ppt@rev.pol.cat.sspro.p1.replay-ppi-uniform@1.0.delta\endcsname{0.0}
\expandafter\def\csname val@rev.pol.cat.sspro.p1.replay-ppi-uniform@1.0.mean\endcsname{0.859}
\expandafter\def\csname cnt@rev.pol.cat.sspro.p1.replay-ppi-uniform@1.0.mean\endcsname{1}
\expandafter\def\csname pct@rev.pol.cat.sspro.p1.replay-ppi-uniform@1.0.mean\endcsname{85.9}
\expandafter\def\csname ppt@rev.pol.cat.sspro.p1.replay-ppi-uniform@1.0.mean\endcsname{$+$85.9}
\expandafter\def\csname val@rev.pol.cat.sspro.p1.replay-ppi-uniform@1.0.mean20\endcsname{0.859}
\expandafter\def\csname cnt@rev.pol.cat.sspro.p1.replay-ppi-uniform@1.0.mean20\endcsname{1}
\expandafter\def\csname pct@rev.pol.cat.sspro.p1.replay-ppi-uniform@1.0.mean20\endcsname{85.9}
\expandafter\def\csname ppt@rev.pol.cat.sspro.p1.replay-ppi-uniform@1.0.mean20\endcsname{$+$85.9}
\expandafter\def\csname val@rev.pol.cat.sspro.p1.replay-ppi-uniform@1.0.sd\endcsname{0.000}
\expandafter\def\csname cnt@rev.pol.cat.sspro.p1.replay-ppi-uniform@1.0.sd\endcsname{0}
\expandafter\def\csname pct@rev.pol.cat.sspro.p1.replay-ppi-uniform@1.0.sd\endcsname{0.0}
\expandafter\def\csname ppt@rev.pol.cat.sspro.p1.replay-ppi-uniform@1.0.sd\endcsname{0.0}
\expandafter\def\csname val@rev.pol.cat.sspro.p1.replay-ppi-uniform@1.0.sd20\endcsname{0.000}
\expandafter\def\csname cnt@rev.pol.cat.sspro.p1.replay-ppi-uniform@1.0.sd20\endcsname{0}
\expandafter\def\csname pct@rev.pol.cat.sspro.p1.replay-ppi-uniform@1.0.sd20\endcsname{0.0}
\expandafter\def\csname ppt@rev.pol.cat.sspro.p1.replay-ppi-uniform@1.0.sd20\endcsname{0.0}
\expandafter\def\csname val@rev.pol.cat.sspro.p1.uniform-fresh@0.1.delta\endcsname{0.001}
\expandafter\def\csname cnt@rev.pol.cat.sspro.p1.uniform-fresh@0.1.delta\endcsname{0}
\expandafter\def\csname pct@rev.pol.cat.sspro.p1.uniform-fresh@0.1.delta\endcsname{0.1}
\expandafter\def\csname ppt@rev.pol.cat.sspro.p1.uniform-fresh@0.1.delta\endcsname{$+$0.1}
\expandafter\def\csname val@rev.pol.cat.sspro.p1.uniform-fresh@0.1.mean\endcsname{0.849}
\expandafter\def\csname cnt@rev.pol.cat.sspro.p1.uniform-fresh@0.1.mean\endcsname{1}
\expandafter\def\csname pct@rev.pol.cat.sspro.p1.uniform-fresh@0.1.mean\endcsname{84.9}
\expandafter\def\csname ppt@rev.pol.cat.sspro.p1.uniform-fresh@0.1.mean\endcsname{$+$84.9}
\expandafter\def\csname val@rev.pol.cat.sspro.p1.uniform-fresh@0.1.mean20\endcsname{0.848}
\expandafter\def\csname cnt@rev.pol.cat.sspro.p1.uniform-fresh@0.1.mean20\endcsname{1}
\expandafter\def\csname pct@rev.pol.cat.sspro.p1.uniform-fresh@0.1.mean20\endcsname{84.8}
\expandafter\def\csname ppt@rev.pol.cat.sspro.p1.uniform-fresh@0.1.mean20\endcsname{$+$84.8}
\expandafter\def\csname val@rev.pol.cat.sspro.p1.uniform-fresh@0.1.sd\endcsname{0.027}
\expandafter\def\csname cnt@rev.pol.cat.sspro.p1.uniform-fresh@0.1.sd\endcsname{0}
\expandafter\def\csname pct@rev.pol.cat.sspro.p1.uniform-fresh@0.1.sd\endcsname{2.7}
\expandafter\def\csname ppt@rev.pol.cat.sspro.p1.uniform-fresh@0.1.sd\endcsname{$+$2.7}
\expandafter\def\csname val@rev.pol.cat.sspro.p1.uniform-fresh@0.1.sd20\endcsname{0.029}
\expandafter\def\csname cnt@rev.pol.cat.sspro.p1.uniform-fresh@0.1.sd20\endcsname{0}
\expandafter\def\csname pct@rev.pol.cat.sspro.p1.uniform-fresh@0.1.sd20\endcsname{2.9}
\expandafter\def\csname ppt@rev.pol.cat.sspro.p1.uniform-fresh@0.1.sd20\endcsname{$+$2.9}
\expandafter\def\csname val@rev.pol.cat.sspro.p1.uniform-fresh@0.25.delta\endcsname{0.003}
\expandafter\def\csname cnt@rev.pol.cat.sspro.p1.uniform-fresh@0.25.delta\endcsname{0}
\expandafter\def\csname pct@rev.pol.cat.sspro.p1.uniform-fresh@0.25.delta\endcsname{0.3}
\expandafter\def\csname ppt@rev.pol.cat.sspro.p1.uniform-fresh@0.25.delta\endcsname{$+$0.3}
\expandafter\def\csname val@rev.pol.cat.sspro.p1.uniform-fresh@0.25.mean\endcsname{0.853}
\expandafter\def\csname cnt@rev.pol.cat.sspro.p1.uniform-fresh@0.25.mean\endcsname{1}
\expandafter\def\csname pct@rev.pol.cat.sspro.p1.uniform-fresh@0.25.mean\endcsname{85.3}
\expandafter\def\csname ppt@rev.pol.cat.sspro.p1.uniform-fresh@0.25.mean\endcsname{$+$85.3}
\expandafter\def\csname val@rev.pol.cat.sspro.p1.uniform-fresh@0.25.mean20\endcsname{0.850}
\expandafter\def\csname cnt@rev.pol.cat.sspro.p1.uniform-fresh@0.25.mean20\endcsname{1}
\expandafter\def\csname pct@rev.pol.cat.sspro.p1.uniform-fresh@0.25.mean20\endcsname{85.0}
\expandafter\def\csname ppt@rev.pol.cat.sspro.p1.uniform-fresh@0.25.mean20\endcsname{$+$85.0}
\expandafter\def\csname val@rev.pol.cat.sspro.p1.uniform-fresh@0.25.sd\endcsname{0.020}
\expandafter\def\csname cnt@rev.pol.cat.sspro.p1.uniform-fresh@0.25.sd\endcsname{0}
\expandafter\def\csname pct@rev.pol.cat.sspro.p1.uniform-fresh@0.25.sd\endcsname{2.0}
\expandafter\def\csname ppt@rev.pol.cat.sspro.p1.uniform-fresh@0.25.sd\endcsname{$+$2.0}
\expandafter\def\csname val@rev.pol.cat.sspro.p1.uniform-fresh@0.25.sd20\endcsname{0.022}
\expandafter\def\csname cnt@rev.pol.cat.sspro.p1.uniform-fresh@0.25.sd20\endcsname{0}
\expandafter\def\csname pct@rev.pol.cat.sspro.p1.uniform-fresh@0.25.sd20\endcsname{2.2}
\expandafter\def\csname ppt@rev.pol.cat.sspro.p1.uniform-fresh@0.25.sd20\endcsname{$+$2.2}
\expandafter\def\csname val@rev.pol.cat.sspro.p1.uniform-fresh@0.5.delta\endcsname{-0.001}
\expandafter\def\csname cnt@rev.pol.cat.sspro.p1.uniform-fresh@0.5.delta\endcsname{0}
\expandafter\def\csname pct@rev.pol.cat.sspro.p1.uniform-fresh@0.5.delta\endcsname{$-$0.1}
\expandafter\def\csname ppt@rev.pol.cat.sspro.p1.uniform-fresh@0.5.delta\endcsname{$-$0.1}
\expandafter\def\csname val@rev.pol.cat.sspro.p1.uniform-fresh@0.5.mean\endcsname{0.858}
\expandafter\def\csname cnt@rev.pol.cat.sspro.p1.uniform-fresh@0.5.mean\endcsname{1}
\expandafter\def\csname pct@rev.pol.cat.sspro.p1.uniform-fresh@0.5.mean\endcsname{85.8}
\expandafter\def\csname ppt@rev.pol.cat.sspro.p1.uniform-fresh@0.5.mean\endcsname{$+$85.8}
\expandafter\def\csname val@rev.pol.cat.sspro.p1.uniform-fresh@0.5.mean20\endcsname{0.859}
\expandafter\def\csname cnt@rev.pol.cat.sspro.p1.uniform-fresh@0.5.mean20\endcsname{1}
\expandafter\def\csname pct@rev.pol.cat.sspro.p1.uniform-fresh@0.5.mean20\endcsname{85.9}
\expandafter\def\csname ppt@rev.pol.cat.sspro.p1.uniform-fresh@0.5.mean20\endcsname{$+$85.9}
\expandafter\def\csname val@rev.pol.cat.sspro.p1.uniform-fresh@0.5.sd\endcsname{0.007}
\expandafter\def\csname cnt@rev.pol.cat.sspro.p1.uniform-fresh@0.5.sd\endcsname{0}
\expandafter\def\csname pct@rev.pol.cat.sspro.p1.uniform-fresh@0.5.sd\endcsname{0.7}
\expandafter\def\csname ppt@rev.pol.cat.sspro.p1.uniform-fresh@0.5.sd\endcsname{$+$0.7}
\expandafter\def\csname val@rev.pol.cat.sspro.p1.uniform-fresh@0.5.sd20\endcsname{0.000}
\expandafter\def\csname cnt@rev.pol.cat.sspro.p1.uniform-fresh@0.5.sd20\endcsname{0}
\expandafter\def\csname pct@rev.pol.cat.sspro.p1.uniform-fresh@0.5.sd20\endcsname{0.0}
\expandafter\def\csname ppt@rev.pol.cat.sspro.p1.uniform-fresh@0.5.sd20\endcsname{0.0}
\expandafter\def\csname val@rev.pol.cat.sspro.p1.uniform-fresh@1.0.delta\endcsname{0.000}
\expandafter\def\csname cnt@rev.pol.cat.sspro.p1.uniform-fresh@1.0.delta\endcsname{0}
\expandafter\def\csname pct@rev.pol.cat.sspro.p1.uniform-fresh@1.0.delta\endcsname{0.0}
\expandafter\def\csname ppt@rev.pol.cat.sspro.p1.uniform-fresh@1.0.delta\endcsname{0.0}
\expandafter\def\csname val@rev.pol.cat.sspro.p1.uniform-fresh@1.0.mean\endcsname{0.859}
\expandafter\def\csname cnt@rev.pol.cat.sspro.p1.uniform-fresh@1.0.mean\endcsname{1}
\expandafter\def\csname pct@rev.pol.cat.sspro.p1.uniform-fresh@1.0.mean\endcsname{85.9}
\expandafter\def\csname ppt@rev.pol.cat.sspro.p1.uniform-fresh@1.0.mean\endcsname{$+$85.9}
\expandafter\def\csname val@rev.pol.cat.sspro.p1.uniform-fresh@1.0.mean20\endcsname{0.859}
\expandafter\def\csname cnt@rev.pol.cat.sspro.p1.uniform-fresh@1.0.mean20\endcsname{1}
\expandafter\def\csname pct@rev.pol.cat.sspro.p1.uniform-fresh@1.0.mean20\endcsname{85.9}
\expandafter\def\csname ppt@rev.pol.cat.sspro.p1.uniform-fresh@1.0.mean20\endcsname{$+$85.9}
\expandafter\def\csname val@rev.pol.cat.sspro.p1.uniform-fresh@1.0.sd\endcsname{0.000}
\expandafter\def\csname cnt@rev.pol.cat.sspro.p1.uniform-fresh@1.0.sd\endcsname{0}
\expandafter\def\csname pct@rev.pol.cat.sspro.p1.uniform-fresh@1.0.sd\endcsname{0.0}
\expandafter\def\csname ppt@rev.pol.cat.sspro.p1.uniform-fresh@1.0.sd\endcsname{0.0}
\expandafter\def\csname val@rev.pol.cat.sspro.p1.uniform-fresh@1.0.sd20\endcsname{0.000}
\expandafter\def\csname cnt@rev.pol.cat.sspro.p1.uniform-fresh@1.0.sd20\endcsname{0}
\expandafter\def\csname pct@rev.pol.cat.sspro.p1.uniform-fresh@1.0.sd20\endcsname{0.0}
\expandafter\def\csname ppt@rev.pol.cat.sspro.p1.uniform-fresh@1.0.sd20\endcsname{0.0}
\expandafter\def\csname val@rev.pol.cat.sspro.p2.best-single@0.1.delta\endcsname{-0.007}
\expandafter\def\csname cnt@rev.pol.cat.sspro.p2.best-single@0.1.delta\endcsname{0}
\expandafter\def\csname pct@rev.pol.cat.sspro.p2.best-single@0.1.delta\endcsname{$-$0.7}
\expandafter\def\csname ppt@rev.pol.cat.sspro.p2.best-single@0.1.delta\endcsname{$-$0.7}
\expandafter\def\csname val@rev.pol.cat.sspro.p2.best-single@0.1.mean\endcsname{0.902}
\expandafter\def\csname cnt@rev.pol.cat.sspro.p2.best-single@0.1.mean\endcsname{1}
\expandafter\def\csname pct@rev.pol.cat.sspro.p2.best-single@0.1.mean\endcsname{90.2}
\expandafter\def\csname ppt@rev.pol.cat.sspro.p2.best-single@0.1.mean\endcsname{$+$90.2}
\expandafter\def\csname val@rev.pol.cat.sspro.p2.best-single@0.1.mean20\endcsname{0.908}
\expandafter\def\csname cnt@rev.pol.cat.sspro.p2.best-single@0.1.mean20\endcsname{1}
\expandafter\def\csname pct@rev.pol.cat.sspro.p2.best-single@0.1.mean20\endcsname{90.8}
\expandafter\def\csname ppt@rev.pol.cat.sspro.p2.best-single@0.1.mean20\endcsname{$+$90.8}
\expandafter\def\csname val@rev.pol.cat.sspro.p2.best-single@0.1.sd\endcsname{0.041}
\expandafter\def\csname cnt@rev.pol.cat.sspro.p2.best-single@0.1.sd\endcsname{0}
\expandafter\def\csname pct@rev.pol.cat.sspro.p2.best-single@0.1.sd\endcsname{4.1}
\expandafter\def\csname ppt@rev.pol.cat.sspro.p2.best-single@0.1.sd\endcsname{$+$4.1}
\expandafter\def\csname val@rev.pol.cat.sspro.p2.best-single@0.1.sd20\endcsname{0.034}
\expandafter\def\csname cnt@rev.pol.cat.sspro.p2.best-single@0.1.sd20\endcsname{0}
\expandafter\def\csname pct@rev.pol.cat.sspro.p2.best-single@0.1.sd20\endcsname{3.4}
\expandafter\def\csname ppt@rev.pol.cat.sspro.p2.best-single@0.1.sd20\endcsname{$+$3.4}
\expandafter\def\csname val@rev.pol.cat.sspro.p2.best-single@0.25.delta\endcsname{-0.007}
\expandafter\def\csname cnt@rev.pol.cat.sspro.p2.best-single@0.25.delta\endcsname{0}
\expandafter\def\csname pct@rev.pol.cat.sspro.p2.best-single@0.25.delta\endcsname{$-$0.7}
\expandafter\def\csname ppt@rev.pol.cat.sspro.p2.best-single@0.25.delta\endcsname{$-$0.7}
\expandafter\def\csname val@rev.pol.cat.sspro.p2.best-single@0.25.mean\endcsname{0.906}
\expandafter\def\csname cnt@rev.pol.cat.sspro.p2.best-single@0.25.mean\endcsname{1}
\expandafter\def\csname pct@rev.pol.cat.sspro.p2.best-single@0.25.mean\endcsname{90.6}
\expandafter\def\csname ppt@rev.pol.cat.sspro.p2.best-single@0.25.mean\endcsname{$+$90.6}
\expandafter\def\csname val@rev.pol.cat.sspro.p2.best-single@0.25.mean20\endcsname{0.913}
\expandafter\def\csname cnt@rev.pol.cat.sspro.p2.best-single@0.25.mean20\endcsname{1}
\expandafter\def\csname pct@rev.pol.cat.sspro.p2.best-single@0.25.mean20\endcsname{91.3}
\expandafter\def\csname ppt@rev.pol.cat.sspro.p2.best-single@0.25.mean20\endcsname{$+$91.3}
\expandafter\def\csname val@rev.pol.cat.sspro.p2.best-single@0.25.sd\endcsname{0.026}
\expandafter\def\csname cnt@rev.pol.cat.sspro.p2.best-single@0.25.sd\endcsname{0}
\expandafter\def\csname pct@rev.pol.cat.sspro.p2.best-single@0.25.sd\endcsname{2.6}
\expandafter\def\csname ppt@rev.pol.cat.sspro.p2.best-single@0.25.sd\endcsname{$+$2.6}
\expandafter\def\csname val@rev.pol.cat.sspro.p2.best-single@0.25.sd20\endcsname{0.008}
\expandafter\def\csname cnt@rev.pol.cat.sspro.p2.best-single@0.25.sd20\endcsname{0}
\expandafter\def\csname pct@rev.pol.cat.sspro.p2.best-single@0.25.sd20\endcsname{0.8}
\expandafter\def\csname ppt@rev.pol.cat.sspro.p2.best-single@0.25.sd20\endcsname{$+$0.8}
\expandafter\def\csname val@rev.pol.cat.sspro.p2.best-single@0.5.delta\endcsname{0.001}
\expandafter\def\csname cnt@rev.pol.cat.sspro.p2.best-single@0.5.delta\endcsname{0}
\expandafter\def\csname pct@rev.pol.cat.sspro.p2.best-single@0.5.delta\endcsname{0.1}
\expandafter\def\csname ppt@rev.pol.cat.sspro.p2.best-single@0.5.delta\endcsname{$+$0.1}
\expandafter\def\csname val@rev.pol.cat.sspro.p2.best-single@0.5.mean\endcsname{0.908}
\expandafter\def\csname cnt@rev.pol.cat.sspro.p2.best-single@0.5.mean\endcsname{1}
\expandafter\def\csname pct@rev.pol.cat.sspro.p2.best-single@0.5.mean\endcsname{90.8}
\expandafter\def\csname ppt@rev.pol.cat.sspro.p2.best-single@0.5.mean\endcsname{$+$90.8}
\expandafter\def\csname val@rev.pol.cat.sspro.p2.best-single@0.5.mean20\endcsname{0.907}
\expandafter\def\csname cnt@rev.pol.cat.sspro.p2.best-single@0.5.mean20\endcsname{1}
\expandafter\def\csname pct@rev.pol.cat.sspro.p2.best-single@0.5.mean20\endcsname{90.7}
\expandafter\def\csname ppt@rev.pol.cat.sspro.p2.best-single@0.5.mean20\endcsname{$+$90.7}
\expandafter\def\csname val@rev.pol.cat.sspro.p2.best-single@0.5.sd\endcsname{0.012}
\expandafter\def\csname cnt@rev.pol.cat.sspro.p2.best-single@0.5.sd\endcsname{0}
\expandafter\def\csname pct@rev.pol.cat.sspro.p2.best-single@0.5.sd\endcsname{1.2}
\expandafter\def\csname ppt@rev.pol.cat.sspro.p2.best-single@0.5.sd\endcsname{$+$1.2}
\expandafter\def\csname val@rev.pol.cat.sspro.p2.best-single@0.5.sd20\endcsname{0.013}
\expandafter\def\csname cnt@rev.pol.cat.sspro.p2.best-single@0.5.sd20\endcsname{0}
\expandafter\def\csname pct@rev.pol.cat.sspro.p2.best-single@0.5.sd20\endcsname{1.3}
\expandafter\def\csname ppt@rev.pol.cat.sspro.p2.best-single@0.5.sd20\endcsname{$+$1.3}
\expandafter\def\csname val@rev.pol.cat.sspro.p2.best-single@1.0.delta\endcsname{0.000}
\expandafter\def\csname cnt@rev.pol.cat.sspro.p2.best-single@1.0.delta\endcsname{0}
\expandafter\def\csname pct@rev.pol.cat.sspro.p2.best-single@1.0.delta\endcsname{0.0}
\expandafter\def\csname ppt@rev.pol.cat.sspro.p2.best-single@1.0.delta\endcsname{0.0}
\expandafter\def\csname val@rev.pol.cat.sspro.p2.best-single@1.0.mean\endcsname{0.917}
\expandafter\def\csname cnt@rev.pol.cat.sspro.p2.best-single@1.0.mean\endcsname{1}
\expandafter\def\csname pct@rev.pol.cat.sspro.p2.best-single@1.0.mean\endcsname{91.7}
\expandafter\def\csname ppt@rev.pol.cat.sspro.p2.best-single@1.0.mean\endcsname{$+$91.7}
\expandafter\def\csname val@rev.pol.cat.sspro.p2.best-single@1.0.mean20\endcsname{0.917}
\expandafter\def\csname cnt@rev.pol.cat.sspro.p2.best-single@1.0.mean20\endcsname{1}
\expandafter\def\csname pct@rev.pol.cat.sspro.p2.best-single@1.0.mean20\endcsname{91.7}
\expandafter\def\csname ppt@rev.pol.cat.sspro.p2.best-single@1.0.mean20\endcsname{$+$91.7}
\expandafter\def\csname val@rev.pol.cat.sspro.p2.best-single@1.0.sd\endcsname{0.000}
\expandafter\def\csname cnt@rev.pol.cat.sspro.p2.best-single@1.0.sd\endcsname{0}
\expandafter\def\csname pct@rev.pol.cat.sspro.p2.best-single@1.0.sd\endcsname{0.0}
\expandafter\def\csname ppt@rev.pol.cat.sspro.p2.best-single@1.0.sd\endcsname{0.0}
\expandafter\def\csname val@rev.pol.cat.sspro.p2.best-single@1.0.sd20\endcsname{0.000}
\expandafter\def\csname cnt@rev.pol.cat.sspro.p2.best-single@1.0.sd20\endcsname{0}
\expandafter\def\csname pct@rev.pol.cat.sspro.p2.best-single@1.0.sd20\endcsname{0.0}
\expandafter\def\csname ppt@rev.pol.cat.sspro.p2.best-single@1.0.sd20\endcsname{0.0}
\expandafter\def\csname val@rev.pol.cat.sspro.p2.generic-ppi-source@0.1.delta\endcsname{-0.001}
\expandafter\def\csname cnt@rev.pol.cat.sspro.p2.generic-ppi-source@0.1.delta\endcsname{0}
\expandafter\def\csname pct@rev.pol.cat.sspro.p2.generic-ppi-source@0.1.delta\endcsname{$-$0.1}
\expandafter\def\csname ppt@rev.pol.cat.sspro.p2.generic-ppi-source@0.1.delta\endcsname{$-$0.1}
\expandafter\def\csname val@rev.pol.cat.sspro.p2.generic-ppi-source@0.1.mean\endcsname{0.893}
\expandafter\def\csname cnt@rev.pol.cat.sspro.p2.generic-ppi-source@0.1.mean\endcsname{1}
\expandafter\def\csname pct@rev.pol.cat.sspro.p2.generic-ppi-source@0.1.mean\endcsname{89.3}
\expandafter\def\csname ppt@rev.pol.cat.sspro.p2.generic-ppi-source@0.1.mean\endcsname{$+$89.3}
\expandafter\def\csname val@rev.pol.cat.sspro.p2.generic-ppi-source@0.1.mean20\endcsname{0.895}
\expandafter\def\csname cnt@rev.pol.cat.sspro.p2.generic-ppi-source@0.1.mean20\endcsname{1}
\expandafter\def\csname pct@rev.pol.cat.sspro.p2.generic-ppi-source@0.1.mean20\endcsname{89.5}
\expandafter\def\csname ppt@rev.pol.cat.sspro.p2.generic-ppi-source@0.1.mean20\endcsname{$+$89.5}
\expandafter\def\csname val@rev.pol.cat.sspro.p2.generic-ppi-source@0.1.sd\endcsname{0.020}
\expandafter\def\csname cnt@rev.pol.cat.sspro.p2.generic-ppi-source@0.1.sd\endcsname{0}
\expandafter\def\csname pct@rev.pol.cat.sspro.p2.generic-ppi-source@0.1.sd\endcsname{2.0}
\expandafter\def\csname ppt@rev.pol.cat.sspro.p2.generic-ppi-source@0.1.sd\endcsname{$+$2.0}
\expandafter\def\csname val@rev.pol.cat.sspro.p2.generic-ppi-source@0.1.sd20\endcsname{0.018}
\expandafter\def\csname cnt@rev.pol.cat.sspro.p2.generic-ppi-source@0.1.sd20\endcsname{0}
\expandafter\def\csname pct@rev.pol.cat.sspro.p2.generic-ppi-source@0.1.sd20\endcsname{1.8}
\expandafter\def\csname ppt@rev.pol.cat.sspro.p2.generic-ppi-source@0.1.sd20\endcsname{$+$1.8}
\expandafter\def\csname val@rev.pol.cat.sspro.p2.generic-ppi-source@0.25.delta\endcsname{0.005}
\expandafter\def\csname cnt@rev.pol.cat.sspro.p2.generic-ppi-source@0.25.delta\endcsname{0}
\expandafter\def\csname pct@rev.pol.cat.sspro.p2.generic-ppi-source@0.25.delta\endcsname{0.5}
\expandafter\def\csname ppt@rev.pol.cat.sspro.p2.generic-ppi-source@0.25.delta\endcsname{$+$0.5}
\expandafter\def\csname val@rev.pol.cat.sspro.p2.generic-ppi-source@0.25.mean\endcsname{0.891}
\expandafter\def\csname cnt@rev.pol.cat.sspro.p2.generic-ppi-source@0.25.mean\endcsname{1}
\expandafter\def\csname pct@rev.pol.cat.sspro.p2.generic-ppi-source@0.25.mean\endcsname{89.1}
\expandafter\def\csname ppt@rev.pol.cat.sspro.p2.generic-ppi-source@0.25.mean\endcsname{$+$89.1}
\expandafter\def\csname val@rev.pol.cat.sspro.p2.generic-ppi-source@0.25.mean20\endcsname{0.885}
\expandafter\def\csname cnt@rev.pol.cat.sspro.p2.generic-ppi-source@0.25.mean20\endcsname{1}
\expandafter\def\csname pct@rev.pol.cat.sspro.p2.generic-ppi-source@0.25.mean20\endcsname{88.5}
\expandafter\def\csname ppt@rev.pol.cat.sspro.p2.generic-ppi-source@0.25.mean20\endcsname{$+$88.5}
\expandafter\def\csname val@rev.pol.cat.sspro.p2.generic-ppi-source@0.25.sd\endcsname{0.021}
\expandafter\def\csname cnt@rev.pol.cat.sspro.p2.generic-ppi-source@0.25.sd\endcsname{0}
\expandafter\def\csname pct@rev.pol.cat.sspro.p2.generic-ppi-source@0.25.sd\endcsname{2.1}
\expandafter\def\csname ppt@rev.pol.cat.sspro.p2.generic-ppi-source@0.25.sd\endcsname{$+$2.1}
\expandafter\def\csname val@rev.pol.cat.sspro.p2.generic-ppi-source@0.25.sd20\endcsname{0.019}
\expandafter\def\csname cnt@rev.pol.cat.sspro.p2.generic-ppi-source@0.25.sd20\endcsname{0}
\expandafter\def\csname pct@rev.pol.cat.sspro.p2.generic-ppi-source@0.25.sd20\endcsname{1.9}
\expandafter\def\csname ppt@rev.pol.cat.sspro.p2.generic-ppi-source@0.25.sd20\endcsname{$+$1.9}
\expandafter\def\csname val@rev.pol.cat.sspro.p2.generic-ppi-source@0.5.delta\endcsname{0.003}
\expandafter\def\csname cnt@rev.pol.cat.sspro.p2.generic-ppi-source@0.5.delta\endcsname{0}
\expandafter\def\csname pct@rev.pol.cat.sspro.p2.generic-ppi-source@0.5.delta\endcsname{0.3}
\expandafter\def\csname ppt@rev.pol.cat.sspro.p2.generic-ppi-source@0.5.delta\endcsname{$+$0.3}
\expandafter\def\csname val@rev.pol.cat.sspro.p2.generic-ppi-source@0.5.mean\endcsname{0.902}
\expandafter\def\csname cnt@rev.pol.cat.sspro.p2.generic-ppi-source@0.5.mean\endcsname{1}
\expandafter\def\csname pct@rev.pol.cat.sspro.p2.generic-ppi-source@0.5.mean\endcsname{90.2}
\expandafter\def\csname ppt@rev.pol.cat.sspro.p2.generic-ppi-source@0.5.mean\endcsname{$+$90.2}
\expandafter\def\csname val@rev.pol.cat.sspro.p2.generic-ppi-source@0.5.mean20\endcsname{0.898}
\expandafter\def\csname cnt@rev.pol.cat.sspro.p2.generic-ppi-source@0.5.mean20\endcsname{1}
\expandafter\def\csname pct@rev.pol.cat.sspro.p2.generic-ppi-source@0.5.mean20\endcsname{89.8}
\expandafter\def\csname ppt@rev.pol.cat.sspro.p2.generic-ppi-source@0.5.mean20\endcsname{$+$89.8}
\expandafter\def\csname val@rev.pol.cat.sspro.p2.generic-ppi-source@0.5.sd\endcsname{0.013}
\expandafter\def\csname cnt@rev.pol.cat.sspro.p2.generic-ppi-source@0.5.sd\endcsname{0}
\expandafter\def\csname pct@rev.pol.cat.sspro.p2.generic-ppi-source@0.5.sd\endcsname{1.3}
\expandafter\def\csname ppt@rev.pol.cat.sspro.p2.generic-ppi-source@0.5.sd\endcsname{$+$1.3}
\expandafter\def\csname val@rev.pol.cat.sspro.p2.generic-ppi-source@0.5.sd20\endcsname{0.016}
\expandafter\def\csname cnt@rev.pol.cat.sspro.p2.generic-ppi-source@0.5.sd20\endcsname{0}
\expandafter\def\csname pct@rev.pol.cat.sspro.p2.generic-ppi-source@0.5.sd20\endcsname{1.6}
\expandafter\def\csname ppt@rev.pol.cat.sspro.p2.generic-ppi-source@0.5.sd20\endcsname{$+$1.6}
\expandafter\def\csname val@rev.pol.cat.sspro.p2.generic-ppi-source@1.0.delta\endcsname{0.000}
\expandafter\def\csname cnt@rev.pol.cat.sspro.p2.generic-ppi-source@1.0.delta\endcsname{0}
\expandafter\def\csname pct@rev.pol.cat.sspro.p2.generic-ppi-source@1.0.delta\endcsname{0.0}
\expandafter\def\csname ppt@rev.pol.cat.sspro.p2.generic-ppi-source@1.0.delta\endcsname{0.0}
\expandafter\def\csname val@rev.pol.cat.sspro.p2.generic-ppi-source@1.0.mean\endcsname{0.906}
\expandafter\def\csname cnt@rev.pol.cat.sspro.p2.generic-ppi-source@1.0.mean\endcsname{1}
\expandafter\def\csname pct@rev.pol.cat.sspro.p2.generic-ppi-source@1.0.mean\endcsname{90.6}
\expandafter\def\csname ppt@rev.pol.cat.sspro.p2.generic-ppi-source@1.0.mean\endcsname{$+$90.6}
\expandafter\def\csname val@rev.pol.cat.sspro.p2.generic-ppi-source@1.0.mean20\endcsname{0.906}
\expandafter\def\csname cnt@rev.pol.cat.sspro.p2.generic-ppi-source@1.0.mean20\endcsname{1}
\expandafter\def\csname pct@rev.pol.cat.sspro.p2.generic-ppi-source@1.0.mean20\endcsname{90.6}
\expandafter\def\csname ppt@rev.pol.cat.sspro.p2.generic-ppi-source@1.0.mean20\endcsname{$+$90.6}
\expandafter\def\csname val@rev.pol.cat.sspro.p2.generic-ppi-source@1.0.sd\endcsname{0.000}
\expandafter\def\csname cnt@rev.pol.cat.sspro.p2.generic-ppi-source@1.0.sd\endcsname{0}
\expandafter\def\csname pct@rev.pol.cat.sspro.p2.generic-ppi-source@1.0.sd\endcsname{0.0}
\expandafter\def\csname ppt@rev.pol.cat.sspro.p2.generic-ppi-source@1.0.sd\endcsname{0.0}
\expandafter\def\csname val@rev.pol.cat.sspro.p2.generic-ppi-source@1.0.sd20\endcsname{0.000}
\expandafter\def\csname cnt@rev.pol.cat.sspro.p2.generic-ppi-source@1.0.sd20\endcsname{0}
\expandafter\def\csname pct@rev.pol.cat.sspro.p2.generic-ppi-source@1.0.sd20\endcsname{0.0}
\expandafter\def\csname ppt@rev.pol.cat.sspro.p2.generic-ppi-source@1.0.sd20\endcsname{0.0}
\expandafter\def\csname val@rev.pol.cat.sspro.p2.rcr-stratified@0.1.delta\endcsname{0.000}
\expandafter\def\csname cnt@rev.pol.cat.sspro.p2.rcr-stratified@0.1.delta\endcsname{0}
\expandafter\def\csname pct@rev.pol.cat.sspro.p2.rcr-stratified@0.1.delta\endcsname{0.0}
\expandafter\def\csname ppt@rev.pol.cat.sspro.p2.rcr-stratified@0.1.delta\endcsname{0.0}
\expandafter\def\csname val@rev.pol.cat.sspro.p2.rcr-stratified@0.1.mean\endcsname{0.856}
\expandafter\def\csname cnt@rev.pol.cat.sspro.p2.rcr-stratified@0.1.mean\endcsname{1}
\expandafter\def\csname pct@rev.pol.cat.sspro.p2.rcr-stratified@0.1.mean\endcsname{85.6}
\expandafter\def\csname ppt@rev.pol.cat.sspro.p2.rcr-stratified@0.1.mean\endcsname{$+$85.6}
\expandafter\def\csname val@rev.pol.cat.sspro.p2.rcr-stratified@0.1.mean20\endcsname{0.856}
\expandafter\def\csname cnt@rev.pol.cat.sspro.p2.rcr-stratified@0.1.mean20\endcsname{1}
\expandafter\def\csname pct@rev.pol.cat.sspro.p2.rcr-stratified@0.1.mean20\endcsname{85.6}
\expandafter\def\csname ppt@rev.pol.cat.sspro.p2.rcr-stratified@0.1.mean20\endcsname{$+$85.6}
\expandafter\def\csname val@rev.pol.cat.sspro.p2.rcr-stratified@0.1.sd\endcsname{0.019}
\expandafter\def\csname cnt@rev.pol.cat.sspro.p2.rcr-stratified@0.1.sd\endcsname{0}
\expandafter\def\csname pct@rev.pol.cat.sspro.p2.rcr-stratified@0.1.sd\endcsname{1.9}
\expandafter\def\csname ppt@rev.pol.cat.sspro.p2.rcr-stratified@0.1.sd\endcsname{$+$1.9}
\expandafter\def\csname val@rev.pol.cat.sspro.p2.rcr-stratified@0.1.sd20\endcsname{0.019}
\expandafter\def\csname cnt@rev.pol.cat.sspro.p2.rcr-stratified@0.1.sd20\endcsname{0}
\expandafter\def\csname pct@rev.pol.cat.sspro.p2.rcr-stratified@0.1.sd20\endcsname{1.9}
\expandafter\def\csname ppt@rev.pol.cat.sspro.p2.rcr-stratified@0.1.sd20\endcsname{$+$1.9}
\expandafter\def\csname val@rev.pol.cat.sspro.p2.rcr-stratified@0.25.delta\endcsname{-0.004}
\expandafter\def\csname cnt@rev.pol.cat.sspro.p2.rcr-stratified@0.25.delta\endcsname{0}
\expandafter\def\csname pct@rev.pol.cat.sspro.p2.rcr-stratified@0.25.delta\endcsname{$-$0.4}
\expandafter\def\csname ppt@rev.pol.cat.sspro.p2.rcr-stratified@0.25.delta\endcsname{$-$0.4}
\expandafter\def\csname val@rev.pol.cat.sspro.p2.rcr-stratified@0.25.mean\endcsname{0.891}
\expandafter\def\csname cnt@rev.pol.cat.sspro.p2.rcr-stratified@0.25.mean\endcsname{1}
\expandafter\def\csname pct@rev.pol.cat.sspro.p2.rcr-stratified@0.25.mean\endcsname{89.1}
\expandafter\def\csname ppt@rev.pol.cat.sspro.p2.rcr-stratified@0.25.mean\endcsname{$+$89.1}
\expandafter\def\csname val@rev.pol.cat.sspro.p2.rcr-stratified@0.25.mean20\endcsname{0.895}
\expandafter\def\csname cnt@rev.pol.cat.sspro.p2.rcr-stratified@0.25.mean20\endcsname{1}
\expandafter\def\csname pct@rev.pol.cat.sspro.p2.rcr-stratified@0.25.mean20\endcsname{89.5}
\expandafter\def\csname ppt@rev.pol.cat.sspro.p2.rcr-stratified@0.25.mean20\endcsname{$+$89.5}
\expandafter\def\csname val@rev.pol.cat.sspro.p2.rcr-stratified@0.25.sd\endcsname{0.022}
\expandafter\def\csname cnt@rev.pol.cat.sspro.p2.rcr-stratified@0.25.sd\endcsname{0}
\expandafter\def\csname pct@rev.pol.cat.sspro.p2.rcr-stratified@0.25.sd\endcsname{2.2}
\expandafter\def\csname ppt@rev.pol.cat.sspro.p2.rcr-stratified@0.25.sd\endcsname{$+$2.2}
\expandafter\def\csname val@rev.pol.cat.sspro.p2.rcr-stratified@0.25.sd20\endcsname{0.018}
\expandafter\def\csname cnt@rev.pol.cat.sspro.p2.rcr-stratified@0.25.sd20\endcsname{0}
\expandafter\def\csname pct@rev.pol.cat.sspro.p2.rcr-stratified@0.25.sd20\endcsname{1.8}
\expandafter\def\csname ppt@rev.pol.cat.sspro.p2.rcr-stratified@0.25.sd20\endcsname{$+$1.8}
\expandafter\def\csname val@rev.pol.cat.sspro.p2.rcr-stratified@0.5.delta\endcsname{-0.002}
\expandafter\def\csname cnt@rev.pol.cat.sspro.p2.rcr-stratified@0.5.delta\endcsname{0}
\expandafter\def\csname pct@rev.pol.cat.sspro.p2.rcr-stratified@0.5.delta\endcsname{$-$0.2}
\expandafter\def\csname ppt@rev.pol.cat.sspro.p2.rcr-stratified@0.5.delta\endcsname{$-$0.2}
\expandafter\def\csname val@rev.pol.cat.sspro.p2.rcr-stratified@0.5.mean\endcsname{0.900}
\expandafter\def\csname cnt@rev.pol.cat.sspro.p2.rcr-stratified@0.5.mean\endcsname{1}
\expandafter\def\csname pct@rev.pol.cat.sspro.p2.rcr-stratified@0.5.mean\endcsname{90.0}
\expandafter\def\csname ppt@rev.pol.cat.sspro.p2.rcr-stratified@0.5.mean\endcsname{$+$90.0}
\expandafter\def\csname val@rev.pol.cat.sspro.p2.rcr-stratified@0.5.mean20\endcsname{0.902}
\expandafter\def\csname cnt@rev.pol.cat.sspro.p2.rcr-stratified@0.5.mean20\endcsname{1}
\expandafter\def\csname pct@rev.pol.cat.sspro.p2.rcr-stratified@0.5.mean20\endcsname{90.2}
\expandafter\def\csname ppt@rev.pol.cat.sspro.p2.rcr-stratified@0.5.mean20\endcsname{$+$90.2}
\expandafter\def\csname val@rev.pol.cat.sspro.p2.rcr-stratified@0.5.sd\endcsname{0.014}
\expandafter\def\csname cnt@rev.pol.cat.sspro.p2.rcr-stratified@0.5.sd\endcsname{0}
\expandafter\def\csname pct@rev.pol.cat.sspro.p2.rcr-stratified@0.5.sd\endcsname{1.4}
\expandafter\def\csname ppt@rev.pol.cat.sspro.p2.rcr-stratified@0.5.sd\endcsname{$+$1.4}
\expandafter\def\csname val@rev.pol.cat.sspro.p2.rcr-stratified@0.5.sd20\endcsname{0.012}
\expandafter\def\csname cnt@rev.pol.cat.sspro.p2.rcr-stratified@0.5.sd20\endcsname{0}
\expandafter\def\csname pct@rev.pol.cat.sspro.p2.rcr-stratified@0.5.sd20\endcsname{1.2}
\expandafter\def\csname ppt@rev.pol.cat.sspro.p2.rcr-stratified@0.5.sd20\endcsname{$+$1.2}
\expandafter\def\csname val@rev.pol.cat.sspro.p2.rcr-stratified@1.0.delta\endcsname{0.000}
\expandafter\def\csname cnt@rev.pol.cat.sspro.p2.rcr-stratified@1.0.delta\endcsname{0}
\expandafter\def\csname pct@rev.pol.cat.sspro.p2.rcr-stratified@1.0.delta\endcsname{0.0}
\expandafter\def\csname ppt@rev.pol.cat.sspro.p2.rcr-stratified@1.0.delta\endcsname{0.0}
\expandafter\def\csname val@rev.pol.cat.sspro.p2.rcr-stratified@1.0.mean\endcsname{0.906}
\expandafter\def\csname cnt@rev.pol.cat.sspro.p2.rcr-stratified@1.0.mean\endcsname{1}
\expandafter\def\csname pct@rev.pol.cat.sspro.p2.rcr-stratified@1.0.mean\endcsname{90.6}
\expandafter\def\csname ppt@rev.pol.cat.sspro.p2.rcr-stratified@1.0.mean\endcsname{$+$90.6}
\expandafter\def\csname val@rev.pol.cat.sspro.p2.rcr-stratified@1.0.mean20\endcsname{0.906}
\expandafter\def\csname cnt@rev.pol.cat.sspro.p2.rcr-stratified@1.0.mean20\endcsname{1}
\expandafter\def\csname pct@rev.pol.cat.sspro.p2.rcr-stratified@1.0.mean20\endcsname{90.6}
\expandafter\def\csname ppt@rev.pol.cat.sspro.p2.rcr-stratified@1.0.mean20\endcsname{$+$90.6}
\expandafter\def\csname val@rev.pol.cat.sspro.p2.rcr-stratified@1.0.sd\endcsname{0.000}
\expandafter\def\csname cnt@rev.pol.cat.sspro.p2.rcr-stratified@1.0.sd\endcsname{0}
\expandafter\def\csname pct@rev.pol.cat.sspro.p2.rcr-stratified@1.0.sd\endcsname{0.0}
\expandafter\def\csname ppt@rev.pol.cat.sspro.p2.rcr-stratified@1.0.sd\endcsname{0.0}
\expandafter\def\csname val@rev.pol.cat.sspro.p2.rcr-stratified@1.0.sd20\endcsname{0.000}
\expandafter\def\csname cnt@rev.pol.cat.sspro.p2.rcr-stratified@1.0.sd20\endcsname{0}
\expandafter\def\csname pct@rev.pol.cat.sspro.p2.rcr-stratified@1.0.sd20\endcsname{0.0}
\expandafter\def\csname ppt@rev.pol.cat.sspro.p2.rcr-stratified@1.0.sd20\endcsname{0.0}
\expandafter\def\csname val@rev.pol.cat.sspro.p2.replay-ppi-uniform@0.1.delta\endcsname{0.001}
\expandafter\def\csname cnt@rev.pol.cat.sspro.p2.replay-ppi-uniform@0.1.delta\endcsname{0}
\expandafter\def\csname pct@rev.pol.cat.sspro.p2.replay-ppi-uniform@0.1.delta\endcsname{0.1}
\expandafter\def\csname ppt@rev.pol.cat.sspro.p2.replay-ppi-uniform@0.1.delta\endcsname{$+$0.1}
\expandafter\def\csname val@rev.pol.cat.sspro.p2.replay-ppi-uniform@0.1.mean\endcsname{0.890}
\expandafter\def\csname cnt@rev.pol.cat.sspro.p2.replay-ppi-uniform@0.1.mean\endcsname{1}
\expandafter\def\csname pct@rev.pol.cat.sspro.p2.replay-ppi-uniform@0.1.mean\endcsname{89.0}
\expandafter\def\csname ppt@rev.pol.cat.sspro.p2.replay-ppi-uniform@0.1.mean\endcsname{$+$89.0}
\expandafter\def\csname val@rev.pol.cat.sspro.p2.replay-ppi-uniform@0.1.mean20\endcsname{0.889}
\expandafter\def\csname cnt@rev.pol.cat.sspro.p2.replay-ppi-uniform@0.1.mean20\endcsname{1}
\expandafter\def\csname pct@rev.pol.cat.sspro.p2.replay-ppi-uniform@0.1.mean20\endcsname{88.9}
\expandafter\def\csname ppt@rev.pol.cat.sspro.p2.replay-ppi-uniform@0.1.mean20\endcsname{$+$88.9}
\expandafter\def\csname val@rev.pol.cat.sspro.p2.replay-ppi-uniform@0.1.sd\endcsname{0.020}
\expandafter\def\csname cnt@rev.pol.cat.sspro.p2.replay-ppi-uniform@0.1.sd\endcsname{0}
\expandafter\def\csname pct@rev.pol.cat.sspro.p2.replay-ppi-uniform@0.1.sd\endcsname{2.0}
\expandafter\def\csname ppt@rev.pol.cat.sspro.p2.replay-ppi-uniform@0.1.sd\endcsname{$+$2.0}
\expandafter\def\csname val@rev.pol.cat.sspro.p2.replay-ppi-uniform@0.1.sd20\endcsname{0.019}
\expandafter\def\csname cnt@rev.pol.cat.sspro.p2.replay-ppi-uniform@0.1.sd20\endcsname{0}
\expandafter\def\csname pct@rev.pol.cat.sspro.p2.replay-ppi-uniform@0.1.sd20\endcsname{1.9}
\expandafter\def\csname ppt@rev.pol.cat.sspro.p2.replay-ppi-uniform@0.1.sd20\endcsname{$+$1.9}
\expandafter\def\csname val@rev.pol.cat.sspro.p2.replay-ppi-uniform@0.25.delta\endcsname{0.005}
\expandafter\def\csname cnt@rev.pol.cat.sspro.p2.replay-ppi-uniform@0.25.delta\endcsname{0}
\expandafter\def\csname pct@rev.pol.cat.sspro.p2.replay-ppi-uniform@0.25.delta\endcsname{0.5}
\expandafter\def\csname ppt@rev.pol.cat.sspro.p2.replay-ppi-uniform@0.25.delta\endcsname{$+$0.5}
\expandafter\def\csname val@rev.pol.cat.sspro.p2.replay-ppi-uniform@0.25.mean\endcsname{0.891}
\expandafter\def\csname cnt@rev.pol.cat.sspro.p2.replay-ppi-uniform@0.25.mean\endcsname{1}
\expandafter\def\csname pct@rev.pol.cat.sspro.p2.replay-ppi-uniform@0.25.mean\endcsname{89.1}
\expandafter\def\csname ppt@rev.pol.cat.sspro.p2.replay-ppi-uniform@0.25.mean\endcsname{$+$89.1}
\expandafter\def\csname val@rev.pol.cat.sspro.p2.replay-ppi-uniform@0.25.mean20\endcsname{0.885}
\expandafter\def\csname cnt@rev.pol.cat.sspro.p2.replay-ppi-uniform@0.25.mean20\endcsname{1}
\expandafter\def\csname pct@rev.pol.cat.sspro.p2.replay-ppi-uniform@0.25.mean20\endcsname{88.5}
\expandafter\def\csname ppt@rev.pol.cat.sspro.p2.replay-ppi-uniform@0.25.mean20\endcsname{$+$88.5}
\expandafter\def\csname val@rev.pol.cat.sspro.p2.replay-ppi-uniform@0.25.sd\endcsname{0.021}
\expandafter\def\csname cnt@rev.pol.cat.sspro.p2.replay-ppi-uniform@0.25.sd\endcsname{0}
\expandafter\def\csname pct@rev.pol.cat.sspro.p2.replay-ppi-uniform@0.25.sd\endcsname{2.1}
\expandafter\def\csname ppt@rev.pol.cat.sspro.p2.replay-ppi-uniform@0.25.sd\endcsname{$+$2.1}
\expandafter\def\csname val@rev.pol.cat.sspro.p2.replay-ppi-uniform@0.25.sd20\endcsname{0.023}
\expandafter\def\csname cnt@rev.pol.cat.sspro.p2.replay-ppi-uniform@0.25.sd20\endcsname{0}
\expandafter\def\csname pct@rev.pol.cat.sspro.p2.replay-ppi-uniform@0.25.sd20\endcsname{2.3}
\expandafter\def\csname ppt@rev.pol.cat.sspro.p2.replay-ppi-uniform@0.25.sd20\endcsname{$+$2.3}
\expandafter\def\csname val@rev.pol.cat.sspro.p2.replay-ppi-uniform@0.5.delta\endcsname{0.005}
\expandafter\def\csname cnt@rev.pol.cat.sspro.p2.replay-ppi-uniform@0.5.delta\endcsname{0}
\expandafter\def\csname pct@rev.pol.cat.sspro.p2.replay-ppi-uniform@0.5.delta\endcsname{0.5}
\expandafter\def\csname ppt@rev.pol.cat.sspro.p2.replay-ppi-uniform@0.5.delta\endcsname{$+$0.5}
\expandafter\def\csname val@rev.pol.cat.sspro.p2.replay-ppi-uniform@0.5.mean\endcsname{0.902}
\expandafter\def\csname cnt@rev.pol.cat.sspro.p2.replay-ppi-uniform@0.5.mean\endcsname{1}
\expandafter\def\csname pct@rev.pol.cat.sspro.p2.replay-ppi-uniform@0.5.mean\endcsname{90.2}
\expandafter\def\csname ppt@rev.pol.cat.sspro.p2.replay-ppi-uniform@0.5.mean\endcsname{$+$90.2}
\expandafter\def\csname val@rev.pol.cat.sspro.p2.replay-ppi-uniform@0.5.mean20\endcsname{0.896}
\expandafter\def\csname cnt@rev.pol.cat.sspro.p2.replay-ppi-uniform@0.5.mean20\endcsname{1}
\expandafter\def\csname pct@rev.pol.cat.sspro.p2.replay-ppi-uniform@0.5.mean20\endcsname{89.6}
\expandafter\def\csname ppt@rev.pol.cat.sspro.p2.replay-ppi-uniform@0.5.mean20\endcsname{$+$89.6}
\expandafter\def\csname val@rev.pol.cat.sspro.p2.replay-ppi-uniform@0.5.sd\endcsname{0.013}
\expandafter\def\csname cnt@rev.pol.cat.sspro.p2.replay-ppi-uniform@0.5.sd\endcsname{0}
\expandafter\def\csname pct@rev.pol.cat.sspro.p2.replay-ppi-uniform@0.5.sd\endcsname{1.3}
\expandafter\def\csname ppt@rev.pol.cat.sspro.p2.replay-ppi-uniform@0.5.sd\endcsname{$+$1.3}
\expandafter\def\csname val@rev.pol.cat.sspro.p2.replay-ppi-uniform@0.5.sd20\endcsname{0.017}
\expandafter\def\csname cnt@rev.pol.cat.sspro.p2.replay-ppi-uniform@0.5.sd20\endcsname{0}
\expandafter\def\csname pct@rev.pol.cat.sspro.p2.replay-ppi-uniform@0.5.sd20\endcsname{1.7}
\expandafter\def\csname ppt@rev.pol.cat.sspro.p2.replay-ppi-uniform@0.5.sd20\endcsname{$+$1.7}
\expandafter\def\csname val@rev.pol.cat.sspro.p2.replay-ppi-uniform@1.0.delta\endcsname{0.000}
\expandafter\def\csname cnt@rev.pol.cat.sspro.p2.replay-ppi-uniform@1.0.delta\endcsname{0}
\expandafter\def\csname pct@rev.pol.cat.sspro.p2.replay-ppi-uniform@1.0.delta\endcsname{0.0}
\expandafter\def\csname ppt@rev.pol.cat.sspro.p2.replay-ppi-uniform@1.0.delta\endcsname{0.0}
\expandafter\def\csname val@rev.pol.cat.sspro.p2.replay-ppi-uniform@1.0.mean\endcsname{0.906}
\expandafter\def\csname cnt@rev.pol.cat.sspro.p2.replay-ppi-uniform@1.0.mean\endcsname{1}
\expandafter\def\csname pct@rev.pol.cat.sspro.p2.replay-ppi-uniform@1.0.mean\endcsname{90.6}
\expandafter\def\csname ppt@rev.pol.cat.sspro.p2.replay-ppi-uniform@1.0.mean\endcsname{$+$90.6}
\expandafter\def\csname val@rev.pol.cat.sspro.p2.replay-ppi-uniform@1.0.mean20\endcsname{0.906}
\expandafter\def\csname cnt@rev.pol.cat.sspro.p2.replay-ppi-uniform@1.0.mean20\endcsname{1}
\expandafter\def\csname pct@rev.pol.cat.sspro.p2.replay-ppi-uniform@1.0.mean20\endcsname{90.6}
\expandafter\def\csname ppt@rev.pol.cat.sspro.p2.replay-ppi-uniform@1.0.mean20\endcsname{$+$90.6}
\expandafter\def\csname val@rev.pol.cat.sspro.p2.replay-ppi-uniform@1.0.sd\endcsname{0.000}
\expandafter\def\csname cnt@rev.pol.cat.sspro.p2.replay-ppi-uniform@1.0.sd\endcsname{0}
\expandafter\def\csname pct@rev.pol.cat.sspro.p2.replay-ppi-uniform@1.0.sd\endcsname{0.0}
\expandafter\def\csname ppt@rev.pol.cat.sspro.p2.replay-ppi-uniform@1.0.sd\endcsname{0.0}
\expandafter\def\csname val@rev.pol.cat.sspro.p2.replay-ppi-uniform@1.0.sd20\endcsname{0.000}
\expandafter\def\csname cnt@rev.pol.cat.sspro.p2.replay-ppi-uniform@1.0.sd20\endcsname{0}
\expandafter\def\csname pct@rev.pol.cat.sspro.p2.replay-ppi-uniform@1.0.sd20\endcsname{0.0}
\expandafter\def\csname ppt@rev.pol.cat.sspro.p2.replay-ppi-uniform@1.0.sd20\endcsname{0.0}
\expandafter\def\csname val@rev.pol.cat.sspro.p2.uniform-fresh@0.1.delta\endcsname{-0.001}
\expandafter\def\csname cnt@rev.pol.cat.sspro.p2.uniform-fresh@0.1.delta\endcsname{0}
\expandafter\def\csname pct@rev.pol.cat.sspro.p2.uniform-fresh@0.1.delta\endcsname{$-$0.1}
\expandafter\def\csname ppt@rev.pol.cat.sspro.p2.uniform-fresh@0.1.delta\endcsname{$-$0.1}
\expandafter\def\csname val@rev.pol.cat.sspro.p2.uniform-fresh@0.1.mean\endcsname{0.894}
\expandafter\def\csname cnt@rev.pol.cat.sspro.p2.uniform-fresh@0.1.mean\endcsname{1}
\expandafter\def\csname pct@rev.pol.cat.sspro.p2.uniform-fresh@0.1.mean\endcsname{89.4}
\expandafter\def\csname ppt@rev.pol.cat.sspro.p2.uniform-fresh@0.1.mean\endcsname{$+$89.4}
\expandafter\def\csname val@rev.pol.cat.sspro.p2.uniform-fresh@0.1.mean20\endcsname{0.895}
\expandafter\def\csname cnt@rev.pol.cat.sspro.p2.uniform-fresh@0.1.mean20\endcsname{1}
\expandafter\def\csname pct@rev.pol.cat.sspro.p2.uniform-fresh@0.1.mean20\endcsname{89.5}
\expandafter\def\csname ppt@rev.pol.cat.sspro.p2.uniform-fresh@0.1.mean20\endcsname{$+$89.5}
\expandafter\def\csname val@rev.pol.cat.sspro.p2.uniform-fresh@0.1.sd\endcsname{0.021}
\expandafter\def\csname cnt@rev.pol.cat.sspro.p2.uniform-fresh@0.1.sd\endcsname{0}
\expandafter\def\csname pct@rev.pol.cat.sspro.p2.uniform-fresh@0.1.sd\endcsname{2.1}
\expandafter\def\csname ppt@rev.pol.cat.sspro.p2.uniform-fresh@0.1.sd\endcsname{$+$2.1}
\expandafter\def\csname val@rev.pol.cat.sspro.p2.uniform-fresh@0.1.sd20\endcsname{0.020}
\expandafter\def\csname cnt@rev.pol.cat.sspro.p2.uniform-fresh@0.1.sd20\endcsname{0}
\expandafter\def\csname pct@rev.pol.cat.sspro.p2.uniform-fresh@0.1.sd20\endcsname{2.0}
\expandafter\def\csname ppt@rev.pol.cat.sspro.p2.uniform-fresh@0.1.sd20\endcsname{$+$2.0}
\expandafter\def\csname val@rev.pol.cat.sspro.p2.uniform-fresh@0.25.delta\endcsname{0.004}
\expandafter\def\csname cnt@rev.pol.cat.sspro.p2.uniform-fresh@0.25.delta\endcsname{0}
\expandafter\def\csname pct@rev.pol.cat.sspro.p2.uniform-fresh@0.25.delta\endcsname{0.4}
\expandafter\def\csname ppt@rev.pol.cat.sspro.p2.uniform-fresh@0.25.delta\endcsname{$+$0.4}
\expandafter\def\csname val@rev.pol.cat.sspro.p2.uniform-fresh@0.25.mean\endcsname{0.895}
\expandafter\def\csname cnt@rev.pol.cat.sspro.p2.uniform-fresh@0.25.mean\endcsname{1}
\expandafter\def\csname pct@rev.pol.cat.sspro.p2.uniform-fresh@0.25.mean\endcsname{89.5}
\expandafter\def\csname ppt@rev.pol.cat.sspro.p2.uniform-fresh@0.25.mean\endcsname{$+$89.5}
\expandafter\def\csname val@rev.pol.cat.sspro.p2.uniform-fresh@0.25.mean20\endcsname{0.891}
\expandafter\def\csname cnt@rev.pol.cat.sspro.p2.uniform-fresh@0.25.mean20\endcsname{1}
\expandafter\def\csname pct@rev.pol.cat.sspro.p2.uniform-fresh@0.25.mean20\endcsname{89.1}
\expandafter\def\csname ppt@rev.pol.cat.sspro.p2.uniform-fresh@0.25.mean20\endcsname{$+$89.1}
\expandafter\def\csname val@rev.pol.cat.sspro.p2.uniform-fresh@0.25.sd\endcsname{0.021}
\expandafter\def\csname cnt@rev.pol.cat.sspro.p2.uniform-fresh@0.25.sd\endcsname{0}
\expandafter\def\csname pct@rev.pol.cat.sspro.p2.uniform-fresh@0.25.sd\endcsname{2.1}
\expandafter\def\csname ppt@rev.pol.cat.sspro.p2.uniform-fresh@0.25.sd\endcsname{$+$2.1}
\expandafter\def\csname val@rev.pol.cat.sspro.p2.uniform-fresh@0.25.sd20\endcsname{0.025}
\expandafter\def\csname cnt@rev.pol.cat.sspro.p2.uniform-fresh@0.25.sd20\endcsname{0}
\expandafter\def\csname pct@rev.pol.cat.sspro.p2.uniform-fresh@0.25.sd20\endcsname{2.5}
\expandafter\def\csname ppt@rev.pol.cat.sspro.p2.uniform-fresh@0.25.sd20\endcsname{$+$2.5}
\expandafter\def\csname val@rev.pol.cat.sspro.p2.uniform-fresh@0.5.delta\endcsname{-0.002}
\expandafter\def\csname cnt@rev.pol.cat.sspro.p2.uniform-fresh@0.5.delta\endcsname{0}
\expandafter\def\csname pct@rev.pol.cat.sspro.p2.uniform-fresh@0.5.delta\endcsname{$-$0.2}
\expandafter\def\csname ppt@rev.pol.cat.sspro.p2.uniform-fresh@0.5.delta\endcsname{$-$0.2}
\expandafter\def\csname val@rev.pol.cat.sspro.p2.uniform-fresh@0.5.mean\endcsname{0.901}
\expandafter\def\csname cnt@rev.pol.cat.sspro.p2.uniform-fresh@0.5.mean\endcsname{1}
\expandafter\def\csname pct@rev.pol.cat.sspro.p2.uniform-fresh@0.5.mean\endcsname{90.1}
\expandafter\def\csname ppt@rev.pol.cat.sspro.p2.uniform-fresh@0.5.mean\endcsname{$+$90.1}
\expandafter\def\csname val@rev.pol.cat.sspro.p2.uniform-fresh@0.5.mean20\endcsname{0.902}
\expandafter\def\csname cnt@rev.pol.cat.sspro.p2.uniform-fresh@0.5.mean20\endcsname{1}
\expandafter\def\csname pct@rev.pol.cat.sspro.p2.uniform-fresh@0.5.mean20\endcsname{90.2}
\expandafter\def\csname ppt@rev.pol.cat.sspro.p2.uniform-fresh@0.5.mean20\endcsname{$+$90.2}
\expandafter\def\csname val@rev.pol.cat.sspro.p2.uniform-fresh@0.5.sd\endcsname{0.015}
\expandafter\def\csname cnt@rev.pol.cat.sspro.p2.uniform-fresh@0.5.sd\endcsname{0}
\expandafter\def\csname pct@rev.pol.cat.sspro.p2.uniform-fresh@0.5.sd\endcsname{1.5}
\expandafter\def\csname ppt@rev.pol.cat.sspro.p2.uniform-fresh@0.5.sd\endcsname{$+$1.5}
\expandafter\def\csname val@rev.pol.cat.sspro.p2.uniform-fresh@0.5.sd20\endcsname{0.012}
\expandafter\def\csname cnt@rev.pol.cat.sspro.p2.uniform-fresh@0.5.sd20\endcsname{0}
\expandafter\def\csname pct@rev.pol.cat.sspro.p2.uniform-fresh@0.5.sd20\endcsname{1.2}
\expandafter\def\csname ppt@rev.pol.cat.sspro.p2.uniform-fresh@0.5.sd20\endcsname{$+$1.2}
\expandafter\def\csname val@rev.pol.cat.sspro.p2.uniform-fresh@1.0.delta\endcsname{0.000}
\expandafter\def\csname cnt@rev.pol.cat.sspro.p2.uniform-fresh@1.0.delta\endcsname{0}
\expandafter\def\csname pct@rev.pol.cat.sspro.p2.uniform-fresh@1.0.delta\endcsname{0.0}
\expandafter\def\csname ppt@rev.pol.cat.sspro.p2.uniform-fresh@1.0.delta\endcsname{0.0}
\expandafter\def\csname val@rev.pol.cat.sspro.p2.uniform-fresh@1.0.mean\endcsname{0.906}
\expandafter\def\csname cnt@rev.pol.cat.sspro.p2.uniform-fresh@1.0.mean\endcsname{1}
\expandafter\def\csname pct@rev.pol.cat.sspro.p2.uniform-fresh@1.0.mean\endcsname{90.6}
\expandafter\def\csname ppt@rev.pol.cat.sspro.p2.uniform-fresh@1.0.mean\endcsname{$+$90.6}
\expandafter\def\csname val@rev.pol.cat.sspro.p2.uniform-fresh@1.0.mean20\endcsname{0.906}
\expandafter\def\csname cnt@rev.pol.cat.sspro.p2.uniform-fresh@1.0.mean20\endcsname{1}
\expandafter\def\csname pct@rev.pol.cat.sspro.p2.uniform-fresh@1.0.mean20\endcsname{90.6}
\expandafter\def\csname ppt@rev.pol.cat.sspro.p2.uniform-fresh@1.0.mean20\endcsname{$+$90.6}
\expandafter\def\csname val@rev.pol.cat.sspro.p2.uniform-fresh@1.0.sd\endcsname{0.000}
\expandafter\def\csname cnt@rev.pol.cat.sspro.p2.uniform-fresh@1.0.sd\endcsname{0}
\expandafter\def\csname pct@rev.pol.cat.sspro.p2.uniform-fresh@1.0.sd\endcsname{0.0}
\expandafter\def\csname ppt@rev.pol.cat.sspro.p2.uniform-fresh@1.0.sd\endcsname{0.0}
\expandafter\def\csname val@rev.pol.cat.sspro.p2.uniform-fresh@1.0.sd20\endcsname{0.000}
\expandafter\def\csname cnt@rev.pol.cat.sspro.p2.uniform-fresh@1.0.sd20\endcsname{0}
\expandafter\def\csname pct@rev.pol.cat.sspro.p2.uniform-fresh@1.0.sd20\endcsname{0.0}
\expandafter\def\csname ppt@rev.pol.cat.sspro.p2.uniform-fresh@1.0.sd20\endcsname{0.0}
\expandafter\def\csname val@rev.pol.cat.uiv.p1.best-single@0.1.delta\endcsname{-0.004}
\expandafter\def\csname cnt@rev.pol.cat.uiv.p1.best-single@0.1.delta\endcsname{0}
\expandafter\def\csname pct@rev.pol.cat.uiv.p1.best-single@0.1.delta\endcsname{$-$0.4}
\expandafter\def\csname ppt@rev.pol.cat.uiv.p1.best-single@0.1.delta\endcsname{$-$0.4}
\expandafter\def\csname val@rev.pol.cat.uiv.p1.best-single@0.1.mean\endcsname{0.786}
\expandafter\def\csname cnt@rev.pol.cat.uiv.p1.best-single@0.1.mean\endcsname{1}
\expandafter\def\csname pct@rev.pol.cat.uiv.p1.best-single@0.1.mean\endcsname{78.6}
\expandafter\def\csname ppt@rev.pol.cat.uiv.p1.best-single@0.1.mean\endcsname{$+$78.6}
\expandafter\def\csname val@rev.pol.cat.uiv.p1.best-single@0.1.mean20\endcsname{0.790}
\expandafter\def\csname cnt@rev.pol.cat.uiv.p1.best-single@0.1.mean20\endcsname{1}
\expandafter\def\csname pct@rev.pol.cat.uiv.p1.best-single@0.1.mean20\endcsname{79.0}
\expandafter\def\csname ppt@rev.pol.cat.uiv.p1.best-single@0.1.mean20\endcsname{$+$79.0}
\expandafter\def\csname val@rev.pol.cat.uiv.p1.best-single@0.1.sd\endcsname{0.034}
\expandafter\def\csname cnt@rev.pol.cat.uiv.p1.best-single@0.1.sd\endcsname{0}
\expandafter\def\csname pct@rev.pol.cat.uiv.p1.best-single@0.1.sd\endcsname{3.4}
\expandafter\def\csname ppt@rev.pol.cat.uiv.p1.best-single@0.1.sd\endcsname{$+$3.4}
\expandafter\def\csname val@rev.pol.cat.uiv.p1.best-single@0.1.sd20\endcsname{0.039}
\expandafter\def\csname cnt@rev.pol.cat.uiv.p1.best-single@0.1.sd20\endcsname{0}
\expandafter\def\csname pct@rev.pol.cat.uiv.p1.best-single@0.1.sd20\endcsname{3.9}
\expandafter\def\csname ppt@rev.pol.cat.uiv.p1.best-single@0.1.sd20\endcsname{$+$3.9}
\expandafter\def\csname val@rev.pol.cat.uiv.p1.best-single@0.25.delta\endcsname{-0.012}
\expandafter\def\csname cnt@rev.pol.cat.uiv.p1.best-single@0.25.delta\endcsname{0}
\expandafter\def\csname pct@rev.pol.cat.uiv.p1.best-single@0.25.delta\endcsname{$-$1.2}
\expandafter\def\csname ppt@rev.pol.cat.uiv.p1.best-single@0.25.delta\endcsname{$-$1.2}
\expandafter\def\csname val@rev.pol.cat.uiv.p1.best-single@0.25.mean\endcsname{0.790}
\expandafter\def\csname cnt@rev.pol.cat.uiv.p1.best-single@0.25.mean\endcsname{1}
\expandafter\def\csname pct@rev.pol.cat.uiv.p1.best-single@0.25.mean\endcsname{79.0}
\expandafter\def\csname ppt@rev.pol.cat.uiv.p1.best-single@0.25.mean\endcsname{$+$79.0}
\expandafter\def\csname val@rev.pol.cat.uiv.p1.best-single@0.25.mean20\endcsname{0.802}
\expandafter\def\csname cnt@rev.pol.cat.uiv.p1.best-single@0.25.mean20\endcsname{1}
\expandafter\def\csname pct@rev.pol.cat.uiv.p1.best-single@0.25.mean20\endcsname{80.2}
\expandafter\def\csname ppt@rev.pol.cat.uiv.p1.best-single@0.25.mean20\endcsname{$+$80.2}
\expandafter\def\csname val@rev.pol.cat.uiv.p1.best-single@0.25.sd\endcsname{0.037}
\expandafter\def\csname cnt@rev.pol.cat.uiv.p1.best-single@0.25.sd\endcsname{0}
\expandafter\def\csname pct@rev.pol.cat.uiv.p1.best-single@0.25.sd\endcsname{3.7}
\expandafter\def\csname ppt@rev.pol.cat.uiv.p1.best-single@0.25.sd\endcsname{$+$3.7}
\expandafter\def\csname val@rev.pol.cat.uiv.p1.best-single@0.25.sd20\endcsname{0.033}
\expandafter\def\csname cnt@rev.pol.cat.uiv.p1.best-single@0.25.sd20\endcsname{0}
\expandafter\def\csname pct@rev.pol.cat.uiv.p1.best-single@0.25.sd20\endcsname{3.3}
\expandafter\def\csname ppt@rev.pol.cat.uiv.p1.best-single@0.25.sd20\endcsname{$+$3.3}
\expandafter\def\csname val@rev.pol.cat.uiv.p1.best-single@0.5.delta\endcsname{-0.014}
\expandafter\def\csname cnt@rev.pol.cat.uiv.p1.best-single@0.5.delta\endcsname{0}
\expandafter\def\csname pct@rev.pol.cat.uiv.p1.best-single@0.5.delta\endcsname{$-$1.4}
\expandafter\def\csname ppt@rev.pol.cat.uiv.p1.best-single@0.5.delta\endcsname{$-$1.4}
\expandafter\def\csname val@rev.pol.cat.uiv.p1.best-single@0.5.mean\endcsname{0.800}
\expandafter\def\csname cnt@rev.pol.cat.uiv.p1.best-single@0.5.mean\endcsname{1}
\expandafter\def\csname pct@rev.pol.cat.uiv.p1.best-single@0.5.mean\endcsname{80.0}
\expandafter\def\csname ppt@rev.pol.cat.uiv.p1.best-single@0.5.mean\endcsname{$+$80.0}
\expandafter\def\csname val@rev.pol.cat.uiv.p1.best-single@0.5.mean20\endcsname{0.814}
\expandafter\def\csname cnt@rev.pol.cat.uiv.p1.best-single@0.5.mean20\endcsname{1}
\expandafter\def\csname pct@rev.pol.cat.uiv.p1.best-single@0.5.mean20\endcsname{81.4}
\expandafter\def\csname ppt@rev.pol.cat.uiv.p1.best-single@0.5.mean20\endcsname{$+$81.4}
\expandafter\def\csname val@rev.pol.cat.uiv.p1.best-single@0.5.sd\endcsname{0.037}
\expandafter\def\csname cnt@rev.pol.cat.uiv.p1.best-single@0.5.sd\endcsname{0}
\expandafter\def\csname pct@rev.pol.cat.uiv.p1.best-single@0.5.sd\endcsname{3.7}
\expandafter\def\csname ppt@rev.pol.cat.uiv.p1.best-single@0.5.sd\endcsname{$+$3.7}
\expandafter\def\csname val@rev.pol.cat.uiv.p1.best-single@0.5.sd20\endcsname{0.031}
\expandafter\def\csname cnt@rev.pol.cat.uiv.p1.best-single@0.5.sd20\endcsname{0}
\expandafter\def\csname pct@rev.pol.cat.uiv.p1.best-single@0.5.sd20\endcsname{3.1}
\expandafter\def\csname ppt@rev.pol.cat.uiv.p1.best-single@0.5.sd20\endcsname{$+$3.1}
\expandafter\def\csname val@rev.pol.cat.uiv.p1.best-single@1.0.delta\endcsname{0.000}
\expandafter\def\csname cnt@rev.pol.cat.uiv.p1.best-single@1.0.delta\endcsname{0}
\expandafter\def\csname pct@rev.pol.cat.uiv.p1.best-single@1.0.delta\endcsname{0.0}
\expandafter\def\csname ppt@rev.pol.cat.uiv.p1.best-single@1.0.delta\endcsname{0.0}
\expandafter\def\csname val@rev.pol.cat.uiv.p1.best-single@1.0.mean\endcsname{0.833}
\expandafter\def\csname cnt@rev.pol.cat.uiv.p1.best-single@1.0.mean\endcsname{1}
\expandafter\def\csname pct@rev.pol.cat.uiv.p1.best-single@1.0.mean\endcsname{83.3}
\expandafter\def\csname ppt@rev.pol.cat.uiv.p1.best-single@1.0.mean\endcsname{$+$83.3}
\expandafter\def\csname val@rev.pol.cat.uiv.p1.best-single@1.0.mean20\endcsname{0.833}
\expandafter\def\csname cnt@rev.pol.cat.uiv.p1.best-single@1.0.mean20\endcsname{1}
\expandafter\def\csname pct@rev.pol.cat.uiv.p1.best-single@1.0.mean20\endcsname{83.3}
\expandafter\def\csname ppt@rev.pol.cat.uiv.p1.best-single@1.0.mean20\endcsname{$+$83.3}
\expandafter\def\csname val@rev.pol.cat.uiv.p1.best-single@1.0.sd\endcsname{0.000}
\expandafter\def\csname cnt@rev.pol.cat.uiv.p1.best-single@1.0.sd\endcsname{0}
\expandafter\def\csname pct@rev.pol.cat.uiv.p1.best-single@1.0.sd\endcsname{0.0}
\expandafter\def\csname ppt@rev.pol.cat.uiv.p1.best-single@1.0.sd\endcsname{0.0}
\expandafter\def\csname val@rev.pol.cat.uiv.p1.best-single@1.0.sd20\endcsname{0.000}
\expandafter\def\csname cnt@rev.pol.cat.uiv.p1.best-single@1.0.sd20\endcsname{0}
\expandafter\def\csname pct@rev.pol.cat.uiv.p1.best-single@1.0.sd20\endcsname{0.0}
\expandafter\def\csname ppt@rev.pol.cat.uiv.p1.best-single@1.0.sd20\endcsname{0.0}
\expandafter\def\csname val@rev.pol.cat.uiv.p1.generic-ppi-source@0.1.delta\endcsname{0.003}
\expandafter\def\csname cnt@rev.pol.cat.uiv.p1.generic-ppi-source@0.1.delta\endcsname{0}
\expandafter\def\csname pct@rev.pol.cat.uiv.p1.generic-ppi-source@0.1.delta\endcsname{0.3}
\expandafter\def\csname ppt@rev.pol.cat.uiv.p1.generic-ppi-source@0.1.delta\endcsname{$+$0.3}
\expandafter\def\csname val@rev.pol.cat.uiv.p1.generic-ppi-source@0.1.mean\endcsname{0.798}
\expandafter\def\csname cnt@rev.pol.cat.uiv.p1.generic-ppi-source@0.1.mean\endcsname{1}
\expandafter\def\csname pct@rev.pol.cat.uiv.p1.generic-ppi-source@0.1.mean\endcsname{79.8}
\expandafter\def\csname ppt@rev.pol.cat.uiv.p1.generic-ppi-source@0.1.mean\endcsname{$+$79.8}
\expandafter\def\csname val@rev.pol.cat.uiv.p1.generic-ppi-source@0.1.mean20\endcsname{0.795}
\expandafter\def\csname cnt@rev.pol.cat.uiv.p1.generic-ppi-source@0.1.mean20\endcsname{1}
\expandafter\def\csname pct@rev.pol.cat.uiv.p1.generic-ppi-source@0.1.mean20\endcsname{79.5}
\expandafter\def\csname ppt@rev.pol.cat.uiv.p1.generic-ppi-source@0.1.mean20\endcsname{$+$79.5}
\expandafter\def\csname val@rev.pol.cat.uiv.p1.generic-ppi-source@0.1.sd\endcsname{0.024}
\expandafter\def\csname cnt@rev.pol.cat.uiv.p1.generic-ppi-source@0.1.sd\endcsname{0}
\expandafter\def\csname pct@rev.pol.cat.uiv.p1.generic-ppi-source@0.1.sd\endcsname{2.4}
\expandafter\def\csname ppt@rev.pol.cat.uiv.p1.generic-ppi-source@0.1.sd\endcsname{$+$2.4}
\expandafter\def\csname val@rev.pol.cat.uiv.p1.generic-ppi-source@0.1.sd20\endcsname{0.021}
\expandafter\def\csname cnt@rev.pol.cat.uiv.p1.generic-ppi-source@0.1.sd20\endcsname{0}
\expandafter\def\csname pct@rev.pol.cat.uiv.p1.generic-ppi-source@0.1.sd20\endcsname{2.1}
\expandafter\def\csname ppt@rev.pol.cat.uiv.p1.generic-ppi-source@0.1.sd20\endcsname{$+$2.1}
\expandafter\def\csname val@rev.pol.cat.uiv.p1.generic-ppi-source@0.25.delta\endcsname{0.007}
\expandafter\def\csname cnt@rev.pol.cat.uiv.p1.generic-ppi-source@0.25.delta\endcsname{0}
\expandafter\def\csname pct@rev.pol.cat.uiv.p1.generic-ppi-source@0.25.delta\endcsname{0.7}
\expandafter\def\csname ppt@rev.pol.cat.uiv.p1.generic-ppi-source@0.25.delta\endcsname{$+$0.7}
\expandafter\def\csname val@rev.pol.cat.uiv.p1.generic-ppi-source@0.25.mean\endcsname{0.797}
\expandafter\def\csname cnt@rev.pol.cat.uiv.p1.generic-ppi-source@0.25.mean\endcsname{1}
\expandafter\def\csname pct@rev.pol.cat.uiv.p1.generic-ppi-source@0.25.mean\endcsname{79.7}
\expandafter\def\csname ppt@rev.pol.cat.uiv.p1.generic-ppi-source@0.25.mean\endcsname{$+$79.7}
\expandafter\def\csname val@rev.pol.cat.uiv.p1.generic-ppi-source@0.25.mean20\endcsname{0.790}
\expandafter\def\csname cnt@rev.pol.cat.uiv.p1.generic-ppi-source@0.25.mean20\endcsname{1}
\expandafter\def\csname pct@rev.pol.cat.uiv.p1.generic-ppi-source@0.25.mean20\endcsname{79.0}
\expandafter\def\csname ppt@rev.pol.cat.uiv.p1.generic-ppi-source@0.25.mean20\endcsname{$+$79.0}
\expandafter\def\csname val@rev.pol.cat.uiv.p1.generic-ppi-source@0.25.sd\endcsname{0.024}
\expandafter\def\csname cnt@rev.pol.cat.uiv.p1.generic-ppi-source@0.25.sd\endcsname{0}
\expandafter\def\csname pct@rev.pol.cat.uiv.p1.generic-ppi-source@0.25.sd\endcsname{2.4}
\expandafter\def\csname ppt@rev.pol.cat.uiv.p1.generic-ppi-source@0.25.sd\endcsname{$+$2.4}
\expandafter\def\csname val@rev.pol.cat.uiv.p1.generic-ppi-source@0.25.sd20\endcsname{0.022}
\expandafter\def\csname cnt@rev.pol.cat.uiv.p1.generic-ppi-source@0.25.sd20\endcsname{0}
\expandafter\def\csname pct@rev.pol.cat.uiv.p1.generic-ppi-source@0.25.sd20\endcsname{2.2}
\expandafter\def\csname ppt@rev.pol.cat.uiv.p1.generic-ppi-source@0.25.sd20\endcsname{$+$2.2}
\expandafter\def\csname val@rev.pol.cat.uiv.p1.generic-ppi-source@0.5.delta\endcsname{-0.001}
\expandafter\def\csname cnt@rev.pol.cat.uiv.p1.generic-ppi-source@0.5.delta\endcsname{0}
\expandafter\def\csname pct@rev.pol.cat.uiv.p1.generic-ppi-source@0.5.delta\endcsname{$-$0.1}
\expandafter\def\csname ppt@rev.pol.cat.uiv.p1.generic-ppi-source@0.5.delta\endcsname{$-$0.1}
\expandafter\def\csname val@rev.pol.cat.uiv.p1.generic-ppi-source@0.5.mean\endcsname{0.799}
\expandafter\def\csname cnt@rev.pol.cat.uiv.p1.generic-ppi-source@0.5.mean\endcsname{1}
\expandafter\def\csname pct@rev.pol.cat.uiv.p1.generic-ppi-source@0.5.mean\endcsname{79.9}
\expandafter\def\csname ppt@rev.pol.cat.uiv.p1.generic-ppi-source@0.5.mean\endcsname{$+$79.9}
\expandafter\def\csname val@rev.pol.cat.uiv.p1.generic-ppi-source@0.5.mean20\endcsname{0.800}
\expandafter\def\csname cnt@rev.pol.cat.uiv.p1.generic-ppi-source@0.5.mean20\endcsname{1}
\expandafter\def\csname pct@rev.pol.cat.uiv.p1.generic-ppi-source@0.5.mean20\endcsname{80.0}
\expandafter\def\csname ppt@rev.pol.cat.uiv.p1.generic-ppi-source@0.5.mean20\endcsname{$+$80.0}
\expandafter\def\csname val@rev.pol.cat.uiv.p1.generic-ppi-source@0.5.sd\endcsname{0.024}
\expandafter\def\csname cnt@rev.pol.cat.uiv.p1.generic-ppi-source@0.5.sd\endcsname{0}
\expandafter\def\csname pct@rev.pol.cat.uiv.p1.generic-ppi-source@0.5.sd\endcsname{2.4}
\expandafter\def\csname ppt@rev.pol.cat.uiv.p1.generic-ppi-source@0.5.sd\endcsname{$+$2.4}
\expandafter\def\csname val@rev.pol.cat.uiv.p1.generic-ppi-source@0.5.sd20\endcsname{0.024}
\expandafter\def\csname cnt@rev.pol.cat.uiv.p1.generic-ppi-source@0.5.sd20\endcsname{0}
\expandafter\def\csname pct@rev.pol.cat.uiv.p1.generic-ppi-source@0.5.sd20\endcsname{2.4}
\expandafter\def\csname ppt@rev.pol.cat.uiv.p1.generic-ppi-source@0.5.sd20\endcsname{$+$2.4}
\expandafter\def\csname val@rev.pol.cat.uiv.p1.generic-ppi-source@1.0.delta\endcsname{0.000}
\expandafter\def\csname cnt@rev.pol.cat.uiv.p1.generic-ppi-source@1.0.delta\endcsname{0}
\expandafter\def\csname pct@rev.pol.cat.uiv.p1.generic-ppi-source@1.0.delta\endcsname{0.0}
\expandafter\def\csname ppt@rev.pol.cat.uiv.p1.generic-ppi-source@1.0.delta\endcsname{0.0}
\expandafter\def\csname val@rev.pol.cat.uiv.p1.generic-ppi-source@1.0.mean\endcsname{0.793}
\expandafter\def\csname cnt@rev.pol.cat.uiv.p1.generic-ppi-source@1.0.mean\endcsname{1}
\expandafter\def\csname pct@rev.pol.cat.uiv.p1.generic-ppi-source@1.0.mean\endcsname{79.3}
\expandafter\def\csname ppt@rev.pol.cat.uiv.p1.generic-ppi-source@1.0.mean\endcsname{$+$79.3}
\expandafter\def\csname val@rev.pol.cat.uiv.p1.generic-ppi-source@1.0.mean20\endcsname{0.793}
\expandafter\def\csname cnt@rev.pol.cat.uiv.p1.generic-ppi-source@1.0.mean20\endcsname{1}
\expandafter\def\csname pct@rev.pol.cat.uiv.p1.generic-ppi-source@1.0.mean20\endcsname{79.3}
\expandafter\def\csname ppt@rev.pol.cat.uiv.p1.generic-ppi-source@1.0.mean20\endcsname{$+$79.3}
\expandafter\def\csname val@rev.pol.cat.uiv.p1.generic-ppi-source@1.0.sd\endcsname{0.000}
\expandafter\def\csname cnt@rev.pol.cat.uiv.p1.generic-ppi-source@1.0.sd\endcsname{0}
\expandafter\def\csname pct@rev.pol.cat.uiv.p1.generic-ppi-source@1.0.sd\endcsname{0.0}
\expandafter\def\csname ppt@rev.pol.cat.uiv.p1.generic-ppi-source@1.0.sd\endcsname{0.0}
\expandafter\def\csname val@rev.pol.cat.uiv.p1.generic-ppi-source@1.0.sd20\endcsname{0.000}
\expandafter\def\csname cnt@rev.pol.cat.uiv.p1.generic-ppi-source@1.0.sd20\endcsname{0}
\expandafter\def\csname pct@rev.pol.cat.uiv.p1.generic-ppi-source@1.0.sd20\endcsname{0.0}
\expandafter\def\csname ppt@rev.pol.cat.uiv.p1.generic-ppi-source@1.0.sd20\endcsname{0.0}
\expandafter\def\csname val@rev.pol.cat.uiv.p1.rcr-stratified@0.1.delta\endcsname{-0.002}
\expandafter\def\csname cnt@rev.pol.cat.uiv.p1.rcr-stratified@0.1.delta\endcsname{0}
\expandafter\def\csname pct@rev.pol.cat.uiv.p1.rcr-stratified@0.1.delta\endcsname{$-$0.2}
\expandafter\def\csname ppt@rev.pol.cat.uiv.p1.rcr-stratified@0.1.delta\endcsname{$-$0.2}
\expandafter\def\csname val@rev.pol.cat.uiv.p1.rcr-stratified@0.1.mean\endcsname{0.805}
\expandafter\def\csname cnt@rev.pol.cat.uiv.p1.rcr-stratified@0.1.mean\endcsname{1}
\expandafter\def\csname pct@rev.pol.cat.uiv.p1.rcr-stratified@0.1.mean\endcsname{80.5}
\expandafter\def\csname ppt@rev.pol.cat.uiv.p1.rcr-stratified@0.1.mean\endcsname{$+$80.5}
\expandafter\def\csname val@rev.pol.cat.uiv.p1.rcr-stratified@0.1.mean20\endcsname{0.807}
\expandafter\def\csname cnt@rev.pol.cat.uiv.p1.rcr-stratified@0.1.mean20\endcsname{1}
\expandafter\def\csname pct@rev.pol.cat.uiv.p1.rcr-stratified@0.1.mean20\endcsname{80.7}
\expandafter\def\csname ppt@rev.pol.cat.uiv.p1.rcr-stratified@0.1.mean20\endcsname{$+$80.7}
\expandafter\def\csname val@rev.pol.cat.uiv.p1.rcr-stratified@0.1.sd\endcsname{0.018}
\expandafter\def\csname cnt@rev.pol.cat.uiv.p1.rcr-stratified@0.1.sd\endcsname{0}
\expandafter\def\csname pct@rev.pol.cat.uiv.p1.rcr-stratified@0.1.sd\endcsname{1.8}
\expandafter\def\csname ppt@rev.pol.cat.uiv.p1.rcr-stratified@0.1.sd\endcsname{$+$1.8}
\expandafter\def\csname val@rev.pol.cat.uiv.p1.rcr-stratified@0.1.sd20\endcsname{0.012}
\expandafter\def\csname cnt@rev.pol.cat.uiv.p1.rcr-stratified@0.1.sd20\endcsname{0}
\expandafter\def\csname pct@rev.pol.cat.uiv.p1.rcr-stratified@0.1.sd20\endcsname{1.2}
\expandafter\def\csname ppt@rev.pol.cat.uiv.p1.rcr-stratified@0.1.sd20\endcsname{$+$1.2}
\expandafter\def\csname val@rev.pol.cat.uiv.p1.rcr-stratified@0.25.delta\endcsname{0.001}
\expandafter\def\csname cnt@rev.pol.cat.uiv.p1.rcr-stratified@0.25.delta\endcsname{0}
\expandafter\def\csname pct@rev.pol.cat.uiv.p1.rcr-stratified@0.25.delta\endcsname{0.1}
\expandafter\def\csname ppt@rev.pol.cat.uiv.p1.rcr-stratified@0.25.delta\endcsname{$+$0.1}
\expandafter\def\csname val@rev.pol.cat.uiv.p1.rcr-stratified@0.25.mean\endcsname{0.805}
\expandafter\def\csname cnt@rev.pol.cat.uiv.p1.rcr-stratified@0.25.mean\endcsname{1}
\expandafter\def\csname pct@rev.pol.cat.uiv.p1.rcr-stratified@0.25.mean\endcsname{80.5}
\expandafter\def\csname ppt@rev.pol.cat.uiv.p1.rcr-stratified@0.25.mean\endcsname{$+$80.5}
\expandafter\def\csname val@rev.pol.cat.uiv.p1.rcr-stratified@0.25.mean20\endcsname{0.804}
\expandafter\def\csname cnt@rev.pol.cat.uiv.p1.rcr-stratified@0.25.mean20\endcsname{1}
\expandafter\def\csname pct@rev.pol.cat.uiv.p1.rcr-stratified@0.25.mean20\endcsname{80.4}
\expandafter\def\csname ppt@rev.pol.cat.uiv.p1.rcr-stratified@0.25.mean20\endcsname{$+$80.4}
\expandafter\def\csname val@rev.pol.cat.uiv.p1.rcr-stratified@0.25.sd\endcsname{0.018}
\expandafter\def\csname cnt@rev.pol.cat.uiv.p1.rcr-stratified@0.25.sd\endcsname{0}
\expandafter\def\csname pct@rev.pol.cat.uiv.p1.rcr-stratified@0.25.sd\endcsname{1.8}
\expandafter\def\csname ppt@rev.pol.cat.uiv.p1.rcr-stratified@0.25.sd\endcsname{$+$1.8}
\expandafter\def\csname val@rev.pol.cat.uiv.p1.rcr-stratified@0.25.sd20\endcsname{0.019}
\expandafter\def\csname cnt@rev.pol.cat.uiv.p1.rcr-stratified@0.25.sd20\endcsname{0}
\expandafter\def\csname pct@rev.pol.cat.uiv.p1.rcr-stratified@0.25.sd20\endcsname{1.9}
\expandafter\def\csname ppt@rev.pol.cat.uiv.p1.rcr-stratified@0.25.sd20\endcsname{$+$1.9}
\expandafter\def\csname val@rev.pol.cat.uiv.p1.rcr-stratified@0.5.delta\endcsname{0.004}
\expandafter\def\csname cnt@rev.pol.cat.uiv.p1.rcr-stratified@0.5.delta\endcsname{0}
\expandafter\def\csname pct@rev.pol.cat.uiv.p1.rcr-stratified@0.5.delta\endcsname{0.4}
\expandafter\def\csname ppt@rev.pol.cat.uiv.p1.rcr-stratified@0.5.delta\endcsname{$+$0.4}
\expandafter\def\csname val@rev.pol.cat.uiv.p1.rcr-stratified@0.5.mean\endcsname{0.804}
\expandafter\def\csname cnt@rev.pol.cat.uiv.p1.rcr-stratified@0.5.mean\endcsname{1}
\expandafter\def\csname pct@rev.pol.cat.uiv.p1.rcr-stratified@0.5.mean\endcsname{80.4}
\expandafter\def\csname ppt@rev.pol.cat.uiv.p1.rcr-stratified@0.5.mean\endcsname{$+$80.4}
\expandafter\def\csname val@rev.pol.cat.uiv.p1.rcr-stratified@0.5.mean20\endcsname{0.799}
\expandafter\def\csname cnt@rev.pol.cat.uiv.p1.rcr-stratified@0.5.mean20\endcsname{1}
\expandafter\def\csname pct@rev.pol.cat.uiv.p1.rcr-stratified@0.5.mean20\endcsname{79.9}
\expandafter\def\csname ppt@rev.pol.cat.uiv.p1.rcr-stratified@0.5.mean20\endcsname{$+$79.9}
\expandafter\def\csname val@rev.pol.cat.uiv.p1.rcr-stratified@0.5.sd\endcsname{0.015}
\expandafter\def\csname cnt@rev.pol.cat.uiv.p1.rcr-stratified@0.5.sd\endcsname{0}
\expandafter\def\csname pct@rev.pol.cat.uiv.p1.rcr-stratified@0.5.sd\endcsname{1.5}
\expandafter\def\csname ppt@rev.pol.cat.uiv.p1.rcr-stratified@0.5.sd\endcsname{$+$1.5}
\expandafter\def\csname val@rev.pol.cat.uiv.p1.rcr-stratified@0.5.sd20\endcsname{0.017}
\expandafter\def\csname cnt@rev.pol.cat.uiv.p1.rcr-stratified@0.5.sd20\endcsname{0}
\expandafter\def\csname pct@rev.pol.cat.uiv.p1.rcr-stratified@0.5.sd20\endcsname{1.7}
\expandafter\def\csname ppt@rev.pol.cat.uiv.p1.rcr-stratified@0.5.sd20\endcsname{$+$1.7}
\expandafter\def\csname val@rev.pol.cat.uiv.p1.rcr-stratified@1.0.delta\endcsname{0.000}
\expandafter\def\csname cnt@rev.pol.cat.uiv.p1.rcr-stratified@1.0.delta\endcsname{0}
\expandafter\def\csname pct@rev.pol.cat.uiv.p1.rcr-stratified@1.0.delta\endcsname{0.0}
\expandafter\def\csname ppt@rev.pol.cat.uiv.p1.rcr-stratified@1.0.delta\endcsname{0.0}
\expandafter\def\csname val@rev.pol.cat.uiv.p1.rcr-stratified@1.0.mean\endcsname{0.793}
\expandafter\def\csname cnt@rev.pol.cat.uiv.p1.rcr-stratified@1.0.mean\endcsname{1}
\expandafter\def\csname pct@rev.pol.cat.uiv.p1.rcr-stratified@1.0.mean\endcsname{79.3}
\expandafter\def\csname ppt@rev.pol.cat.uiv.p1.rcr-stratified@1.0.mean\endcsname{$+$79.3}
\expandafter\def\csname val@rev.pol.cat.uiv.p1.rcr-stratified@1.0.mean20\endcsname{0.793}
\expandafter\def\csname cnt@rev.pol.cat.uiv.p1.rcr-stratified@1.0.mean20\endcsname{1}
\expandafter\def\csname pct@rev.pol.cat.uiv.p1.rcr-stratified@1.0.mean20\endcsname{79.3}
\expandafter\def\csname ppt@rev.pol.cat.uiv.p1.rcr-stratified@1.0.mean20\endcsname{$+$79.3}
\expandafter\def\csname val@rev.pol.cat.uiv.p1.rcr-stratified@1.0.sd\endcsname{0.000}
\expandafter\def\csname cnt@rev.pol.cat.uiv.p1.rcr-stratified@1.0.sd\endcsname{0}
\expandafter\def\csname pct@rev.pol.cat.uiv.p1.rcr-stratified@1.0.sd\endcsname{0.0}
\expandafter\def\csname ppt@rev.pol.cat.uiv.p1.rcr-stratified@1.0.sd\endcsname{0.0}
\expandafter\def\csname val@rev.pol.cat.uiv.p1.rcr-stratified@1.0.sd20\endcsname{0.000}
\expandafter\def\csname cnt@rev.pol.cat.uiv.p1.rcr-stratified@1.0.sd20\endcsname{0}
\expandafter\def\csname pct@rev.pol.cat.uiv.p1.rcr-stratified@1.0.sd20\endcsname{0.0}
\expandafter\def\csname ppt@rev.pol.cat.uiv.p1.rcr-stratified@1.0.sd20\endcsname{0.0}
\expandafter\def\csname val@rev.pol.cat.uiv.p1.replay-ppi-uniform@0.1.delta\endcsname{-0.003}
\expandafter\def\csname cnt@rev.pol.cat.uiv.p1.replay-ppi-uniform@0.1.delta\endcsname{0}
\expandafter\def\csname pct@rev.pol.cat.uiv.p1.replay-ppi-uniform@0.1.delta\endcsname{$-$0.3}
\expandafter\def\csname ppt@rev.pol.cat.uiv.p1.replay-ppi-uniform@0.1.delta\endcsname{$-$0.3}
\expandafter\def\csname val@rev.pol.cat.uiv.p1.replay-ppi-uniform@0.1.mean\endcsname{0.801}
\expandafter\def\csname cnt@rev.pol.cat.uiv.p1.replay-ppi-uniform@0.1.mean\endcsname{1}
\expandafter\def\csname pct@rev.pol.cat.uiv.p1.replay-ppi-uniform@0.1.mean\endcsname{80.1}
\expandafter\def\csname ppt@rev.pol.cat.uiv.p1.replay-ppi-uniform@0.1.mean\endcsname{$+$80.1}
\expandafter\def\csname val@rev.pol.cat.uiv.p1.replay-ppi-uniform@0.1.mean20\endcsname{0.803}
\expandafter\def\csname cnt@rev.pol.cat.uiv.p1.replay-ppi-uniform@0.1.mean20\endcsname{1}
\expandafter\def\csname pct@rev.pol.cat.uiv.p1.replay-ppi-uniform@0.1.mean20\endcsname{80.3}
\expandafter\def\csname ppt@rev.pol.cat.uiv.p1.replay-ppi-uniform@0.1.mean20\endcsname{$+$80.3}
\expandafter\def\csname val@rev.pol.cat.uiv.p1.replay-ppi-uniform@0.1.sd\endcsname{0.022}
\expandafter\def\csname cnt@rev.pol.cat.uiv.p1.replay-ppi-uniform@0.1.sd\endcsname{0}
\expandafter\def\csname pct@rev.pol.cat.uiv.p1.replay-ppi-uniform@0.1.sd\endcsname{2.2}
\expandafter\def\csname ppt@rev.pol.cat.uiv.p1.replay-ppi-uniform@0.1.sd\endcsname{$+$2.2}
\expandafter\def\csname val@rev.pol.cat.uiv.p1.replay-ppi-uniform@0.1.sd20\endcsname{0.022}
\expandafter\def\csname cnt@rev.pol.cat.uiv.p1.replay-ppi-uniform@0.1.sd20\endcsname{0}
\expandafter\def\csname pct@rev.pol.cat.uiv.p1.replay-ppi-uniform@0.1.sd20\endcsname{2.2}
\expandafter\def\csname ppt@rev.pol.cat.uiv.p1.replay-ppi-uniform@0.1.sd20\endcsname{$+$2.2}
\expandafter\def\csname val@rev.pol.cat.uiv.p1.replay-ppi-uniform@0.25.delta\endcsname{-0.002}
\expandafter\def\csname cnt@rev.pol.cat.uiv.p1.replay-ppi-uniform@0.25.delta\endcsname{0}
\expandafter\def\csname pct@rev.pol.cat.uiv.p1.replay-ppi-uniform@0.25.delta\endcsname{$-$0.2}
\expandafter\def\csname ppt@rev.pol.cat.uiv.p1.replay-ppi-uniform@0.25.delta\endcsname{$-$0.2}
\expandafter\def\csname val@rev.pol.cat.uiv.p1.replay-ppi-uniform@0.25.mean\endcsname{0.800}
\expandafter\def\csname cnt@rev.pol.cat.uiv.p1.replay-ppi-uniform@0.25.mean\endcsname{1}
\expandafter\def\csname pct@rev.pol.cat.uiv.p1.replay-ppi-uniform@0.25.mean\endcsname{80.0}
\expandafter\def\csname ppt@rev.pol.cat.uiv.p1.replay-ppi-uniform@0.25.mean\endcsname{$+$80.0}
\expandafter\def\csname val@rev.pol.cat.uiv.p1.replay-ppi-uniform@0.25.mean20\endcsname{0.802}
\expandafter\def\csname cnt@rev.pol.cat.uiv.p1.replay-ppi-uniform@0.25.mean20\endcsname{1}
\expandafter\def\csname pct@rev.pol.cat.uiv.p1.replay-ppi-uniform@0.25.mean20\endcsname{80.2}
\expandafter\def\csname ppt@rev.pol.cat.uiv.p1.replay-ppi-uniform@0.25.mean20\endcsname{$+$80.2}
\expandafter\def\csname val@rev.pol.cat.uiv.p1.replay-ppi-uniform@0.25.sd\endcsname{0.022}
\expandafter\def\csname cnt@rev.pol.cat.uiv.p1.replay-ppi-uniform@0.25.sd\endcsname{0}
\expandafter\def\csname pct@rev.pol.cat.uiv.p1.replay-ppi-uniform@0.25.sd\endcsname{2.2}
\expandafter\def\csname ppt@rev.pol.cat.uiv.p1.replay-ppi-uniform@0.25.sd\endcsname{$+$2.2}
\expandafter\def\csname val@rev.pol.cat.uiv.p1.replay-ppi-uniform@0.25.sd20\endcsname{0.023}
\expandafter\def\csname cnt@rev.pol.cat.uiv.p1.replay-ppi-uniform@0.25.sd20\endcsname{0}
\expandafter\def\csname pct@rev.pol.cat.uiv.p1.replay-ppi-uniform@0.25.sd20\endcsname{2.3}
\expandafter\def\csname ppt@rev.pol.cat.uiv.p1.replay-ppi-uniform@0.25.sd20\endcsname{$+$2.3}
\expandafter\def\csname val@rev.pol.cat.uiv.p1.replay-ppi-uniform@0.5.delta\endcsname{-0.007}
\expandafter\def\csname cnt@rev.pol.cat.uiv.p1.replay-ppi-uniform@0.5.delta\endcsname{0}
\expandafter\def\csname pct@rev.pol.cat.uiv.p1.replay-ppi-uniform@0.5.delta\endcsname{$-$0.7}
\expandafter\def\csname ppt@rev.pol.cat.uiv.p1.replay-ppi-uniform@0.5.delta\endcsname{$-$0.7}
\expandafter\def\csname val@rev.pol.cat.uiv.p1.replay-ppi-uniform@0.5.mean\endcsname{0.803}
\expandafter\def\csname cnt@rev.pol.cat.uiv.p1.replay-ppi-uniform@0.5.mean\endcsname{1}
\expandafter\def\csname pct@rev.pol.cat.uiv.p1.replay-ppi-uniform@0.5.mean\endcsname{80.3}
\expandafter\def\csname ppt@rev.pol.cat.uiv.p1.replay-ppi-uniform@0.5.mean\endcsname{$+$80.3}
\expandafter\def\csname val@rev.pol.cat.uiv.p1.replay-ppi-uniform@0.5.mean20\endcsname{0.810}
\expandafter\def\csname cnt@rev.pol.cat.uiv.p1.replay-ppi-uniform@0.5.mean20\endcsname{1}
\expandafter\def\csname pct@rev.pol.cat.uiv.p1.replay-ppi-uniform@0.5.mean20\endcsname{81.0}
\expandafter\def\csname ppt@rev.pol.cat.uiv.p1.replay-ppi-uniform@0.5.mean20\endcsname{$+$81.0}
\expandafter\def\csname val@rev.pol.cat.uiv.p1.replay-ppi-uniform@0.5.sd\endcsname{0.020}
\expandafter\def\csname cnt@rev.pol.cat.uiv.p1.replay-ppi-uniform@0.5.sd\endcsname{0}
\expandafter\def\csname pct@rev.pol.cat.uiv.p1.replay-ppi-uniform@0.5.sd\endcsname{2.0}
\expandafter\def\csname ppt@rev.pol.cat.uiv.p1.replay-ppi-uniform@0.5.sd\endcsname{$+$2.0}
\expandafter\def\csname val@rev.pol.cat.uiv.p1.replay-ppi-uniform@0.5.sd20\endcsname{0.018}
\expandafter\def\csname cnt@rev.pol.cat.uiv.p1.replay-ppi-uniform@0.5.sd20\endcsname{0}
\expandafter\def\csname pct@rev.pol.cat.uiv.p1.replay-ppi-uniform@0.5.sd20\endcsname{1.8}
\expandafter\def\csname ppt@rev.pol.cat.uiv.p1.replay-ppi-uniform@0.5.sd20\endcsname{$+$1.8}
\expandafter\def\csname val@rev.pol.cat.uiv.p1.replay-ppi-uniform@1.0.delta\endcsname{0.000}
\expandafter\def\csname cnt@rev.pol.cat.uiv.p1.replay-ppi-uniform@1.0.delta\endcsname{0}
\expandafter\def\csname pct@rev.pol.cat.uiv.p1.replay-ppi-uniform@1.0.delta\endcsname{0.0}
\expandafter\def\csname ppt@rev.pol.cat.uiv.p1.replay-ppi-uniform@1.0.delta\endcsname{0.0}
\expandafter\def\csname val@rev.pol.cat.uiv.p1.replay-ppi-uniform@1.0.mean\endcsname{0.793}
\expandafter\def\csname cnt@rev.pol.cat.uiv.p1.replay-ppi-uniform@1.0.mean\endcsname{1}
\expandafter\def\csname pct@rev.pol.cat.uiv.p1.replay-ppi-uniform@1.0.mean\endcsname{79.3}
\expandafter\def\csname ppt@rev.pol.cat.uiv.p1.replay-ppi-uniform@1.0.mean\endcsname{$+$79.3}
\expandafter\def\csname val@rev.pol.cat.uiv.p1.replay-ppi-uniform@1.0.mean20\endcsname{0.793}
\expandafter\def\csname cnt@rev.pol.cat.uiv.p1.replay-ppi-uniform@1.0.mean20\endcsname{1}
\expandafter\def\csname pct@rev.pol.cat.uiv.p1.replay-ppi-uniform@1.0.mean20\endcsname{79.3}
\expandafter\def\csname ppt@rev.pol.cat.uiv.p1.replay-ppi-uniform@1.0.mean20\endcsname{$+$79.3}
\expandafter\def\csname val@rev.pol.cat.uiv.p1.replay-ppi-uniform@1.0.sd\endcsname{0.000}
\expandafter\def\csname cnt@rev.pol.cat.uiv.p1.replay-ppi-uniform@1.0.sd\endcsname{0}
\expandafter\def\csname pct@rev.pol.cat.uiv.p1.replay-ppi-uniform@1.0.sd\endcsname{0.0}
\expandafter\def\csname ppt@rev.pol.cat.uiv.p1.replay-ppi-uniform@1.0.sd\endcsname{0.0}
\expandafter\def\csname val@rev.pol.cat.uiv.p1.replay-ppi-uniform@1.0.sd20\endcsname{0.000}
\expandafter\def\csname cnt@rev.pol.cat.uiv.p1.replay-ppi-uniform@1.0.sd20\endcsname{0}
\expandafter\def\csname pct@rev.pol.cat.uiv.p1.replay-ppi-uniform@1.0.sd20\endcsname{0.0}
\expandafter\def\csname ppt@rev.pol.cat.uiv.p1.replay-ppi-uniform@1.0.sd20\endcsname{0.0}
\expandafter\def\csname val@rev.pol.cat.uiv.p1.uniform-fresh@0.1.delta\endcsname{-0.005}
\expandafter\def\csname cnt@rev.pol.cat.uiv.p1.uniform-fresh@0.1.delta\endcsname{0}
\expandafter\def\csname pct@rev.pol.cat.uiv.p1.uniform-fresh@0.1.delta\endcsname{$-$0.5}
\expandafter\def\csname ppt@rev.pol.cat.uiv.p1.uniform-fresh@0.1.delta\endcsname{$-$0.5}
\expandafter\def\csname val@rev.pol.cat.uiv.p1.uniform-fresh@0.1.mean\endcsname{0.785}
\expandafter\def\csname cnt@rev.pol.cat.uiv.p1.uniform-fresh@0.1.mean\endcsname{1}
\expandafter\def\csname pct@rev.pol.cat.uiv.p1.uniform-fresh@0.1.mean\endcsname{78.5}
\expandafter\def\csname ppt@rev.pol.cat.uiv.p1.uniform-fresh@0.1.mean\endcsname{$+$78.5}
\expandafter\def\csname val@rev.pol.cat.uiv.p1.uniform-fresh@0.1.mean20\endcsname{0.790}
\expandafter\def\csname cnt@rev.pol.cat.uiv.p1.uniform-fresh@0.1.mean20\endcsname{1}
\expandafter\def\csname pct@rev.pol.cat.uiv.p1.uniform-fresh@0.1.mean20\endcsname{79.0}
\expandafter\def\csname ppt@rev.pol.cat.uiv.p1.uniform-fresh@0.1.mean20\endcsname{$+$79.0}
\expandafter\def\csname val@rev.pol.cat.uiv.p1.uniform-fresh@0.1.sd\endcsname{0.026}
\expandafter\def\csname cnt@rev.pol.cat.uiv.p1.uniform-fresh@0.1.sd\endcsname{0}
\expandafter\def\csname pct@rev.pol.cat.uiv.p1.uniform-fresh@0.1.sd\endcsname{2.6}
\expandafter\def\csname ppt@rev.pol.cat.uiv.p1.uniform-fresh@0.1.sd\endcsname{$+$2.6}
\expandafter\def\csname val@rev.pol.cat.uiv.p1.uniform-fresh@0.1.sd20\endcsname{0.027}
\expandafter\def\csname cnt@rev.pol.cat.uiv.p1.uniform-fresh@0.1.sd20\endcsname{0}
\expandafter\def\csname pct@rev.pol.cat.uiv.p1.uniform-fresh@0.1.sd20\endcsname{2.7}
\expandafter\def\csname ppt@rev.pol.cat.uiv.p1.uniform-fresh@0.1.sd20\endcsname{$+$2.7}
\expandafter\def\csname val@rev.pol.cat.uiv.p1.uniform-fresh@0.25.delta\endcsname{-0.005}
\expandafter\def\csname cnt@rev.pol.cat.uiv.p1.uniform-fresh@0.25.delta\endcsname{0}
\expandafter\def\csname pct@rev.pol.cat.uiv.p1.uniform-fresh@0.25.delta\endcsname{$-$0.5}
\expandafter\def\csname ppt@rev.pol.cat.uiv.p1.uniform-fresh@0.25.delta\endcsname{$-$0.5}
\expandafter\def\csname val@rev.pol.cat.uiv.p1.uniform-fresh@0.25.mean\endcsname{0.792}
\expandafter\def\csname cnt@rev.pol.cat.uiv.p1.uniform-fresh@0.25.mean\endcsname{1}
\expandafter\def\csname pct@rev.pol.cat.uiv.p1.uniform-fresh@0.25.mean\endcsname{79.2}
\expandafter\def\csname ppt@rev.pol.cat.uiv.p1.uniform-fresh@0.25.mean\endcsname{$+$79.2}
\expandafter\def\csname val@rev.pol.cat.uiv.p1.uniform-fresh@0.25.mean20\endcsname{0.797}
\expandafter\def\csname cnt@rev.pol.cat.uiv.p1.uniform-fresh@0.25.mean20\endcsname{1}
\expandafter\def\csname pct@rev.pol.cat.uiv.p1.uniform-fresh@0.25.mean20\endcsname{79.7}
\expandafter\def\csname ppt@rev.pol.cat.uiv.p1.uniform-fresh@0.25.mean20\endcsname{$+$79.7}
\expandafter\def\csname val@rev.pol.cat.uiv.p1.uniform-fresh@0.25.sd\endcsname{0.026}
\expandafter\def\csname cnt@rev.pol.cat.uiv.p1.uniform-fresh@0.25.sd\endcsname{0}
\expandafter\def\csname pct@rev.pol.cat.uiv.p1.uniform-fresh@0.25.sd\endcsname{2.6}
\expandafter\def\csname ppt@rev.pol.cat.uiv.p1.uniform-fresh@0.25.sd\endcsname{$+$2.6}
\expandafter\def\csname val@rev.pol.cat.uiv.p1.uniform-fresh@0.25.sd20\endcsname{0.025}
\expandafter\def\csname cnt@rev.pol.cat.uiv.p1.uniform-fresh@0.25.sd20\endcsname{0}
\expandafter\def\csname pct@rev.pol.cat.uiv.p1.uniform-fresh@0.25.sd20\endcsname{2.5}
\expandafter\def\csname ppt@rev.pol.cat.uiv.p1.uniform-fresh@0.25.sd20\endcsname{$+$2.5}
\expandafter\def\csname val@rev.pol.cat.uiv.p1.uniform-fresh@0.5.delta\endcsname{-0.009}
\expandafter\def\csname cnt@rev.pol.cat.uiv.p1.uniform-fresh@0.5.delta\endcsname{0}
\expandafter\def\csname pct@rev.pol.cat.uiv.p1.uniform-fresh@0.5.delta\endcsname{$-$0.9}
\expandafter\def\csname ppt@rev.pol.cat.uiv.p1.uniform-fresh@0.5.delta\endcsname{$-$0.9}
\expandafter\def\csname val@rev.pol.cat.uiv.p1.uniform-fresh@0.5.mean\endcsname{0.796}
\expandafter\def\csname cnt@rev.pol.cat.uiv.p1.uniform-fresh@0.5.mean\endcsname{1}
\expandafter\def\csname pct@rev.pol.cat.uiv.p1.uniform-fresh@0.5.mean\endcsname{79.6}
\expandafter\def\csname ppt@rev.pol.cat.uiv.p1.uniform-fresh@0.5.mean\endcsname{$+$79.6}
\expandafter\def\csname val@rev.pol.cat.uiv.p1.uniform-fresh@0.5.mean20\endcsname{0.806}
\expandafter\def\csname cnt@rev.pol.cat.uiv.p1.uniform-fresh@0.5.mean20\endcsname{1}
\expandafter\def\csname pct@rev.pol.cat.uiv.p1.uniform-fresh@0.5.mean20\endcsname{80.6}
\expandafter\def\csname ppt@rev.pol.cat.uiv.p1.uniform-fresh@0.5.mean20\endcsname{$+$80.6}
\expandafter\def\csname val@rev.pol.cat.uiv.p1.uniform-fresh@0.5.sd\endcsname{0.025}
\expandafter\def\csname cnt@rev.pol.cat.uiv.p1.uniform-fresh@0.5.sd\endcsname{0}
\expandafter\def\csname pct@rev.pol.cat.uiv.p1.uniform-fresh@0.5.sd\endcsname{2.5}
\expandafter\def\csname ppt@rev.pol.cat.uiv.p1.uniform-fresh@0.5.sd\endcsname{$+$2.5}
\expandafter\def\csname val@rev.pol.cat.uiv.p1.uniform-fresh@0.5.sd20\endcsname{0.025}
\expandafter\def\csname cnt@rev.pol.cat.uiv.p1.uniform-fresh@0.5.sd20\endcsname{0}
\expandafter\def\csname pct@rev.pol.cat.uiv.p1.uniform-fresh@0.5.sd20\endcsname{2.5}
\expandafter\def\csname ppt@rev.pol.cat.uiv.p1.uniform-fresh@0.5.sd20\endcsname{$+$2.5}
\expandafter\def\csname val@rev.pol.cat.uiv.p1.uniform-fresh@1.0.delta\endcsname{0.000}
\expandafter\def\csname cnt@rev.pol.cat.uiv.p1.uniform-fresh@1.0.delta\endcsname{0}
\expandafter\def\csname pct@rev.pol.cat.uiv.p1.uniform-fresh@1.0.delta\endcsname{0.0}
\expandafter\def\csname ppt@rev.pol.cat.uiv.p1.uniform-fresh@1.0.delta\endcsname{0.0}
\expandafter\def\csname val@rev.pol.cat.uiv.p1.uniform-fresh@1.0.mean\endcsname{0.793}
\expandafter\def\csname cnt@rev.pol.cat.uiv.p1.uniform-fresh@1.0.mean\endcsname{1}
\expandafter\def\csname pct@rev.pol.cat.uiv.p1.uniform-fresh@1.0.mean\endcsname{79.3}
\expandafter\def\csname ppt@rev.pol.cat.uiv.p1.uniform-fresh@1.0.mean\endcsname{$+$79.3}
\expandafter\def\csname val@rev.pol.cat.uiv.p1.uniform-fresh@1.0.mean20\endcsname{0.793}
\expandafter\def\csname cnt@rev.pol.cat.uiv.p1.uniform-fresh@1.0.mean20\endcsname{1}
\expandafter\def\csname pct@rev.pol.cat.uiv.p1.uniform-fresh@1.0.mean20\endcsname{79.3}
\expandafter\def\csname ppt@rev.pol.cat.uiv.p1.uniform-fresh@1.0.mean20\endcsname{$+$79.3}
\expandafter\def\csname val@rev.pol.cat.uiv.p1.uniform-fresh@1.0.sd\endcsname{0.000}
\expandafter\def\csname cnt@rev.pol.cat.uiv.p1.uniform-fresh@1.0.sd\endcsname{0}
\expandafter\def\csname pct@rev.pol.cat.uiv.p1.uniform-fresh@1.0.sd\endcsname{0.0}
\expandafter\def\csname ppt@rev.pol.cat.uiv.p1.uniform-fresh@1.0.sd\endcsname{0.0}
\expandafter\def\csname val@rev.pol.cat.uiv.p1.uniform-fresh@1.0.sd20\endcsname{0.000}
\expandafter\def\csname cnt@rev.pol.cat.uiv.p1.uniform-fresh@1.0.sd20\endcsname{0}
\expandafter\def\csname pct@rev.pol.cat.uiv.p1.uniform-fresh@1.0.sd20\endcsname{0.0}
\expandafter\def\csname ppt@rev.pol.cat.uiv.p1.uniform-fresh@1.0.sd20\endcsname{0.0}
\expandafter\def\csname val@rev.pol.cat.uiv.p2.best-single@0.1.delta\endcsname{-0.002}
\expandafter\def\csname cnt@rev.pol.cat.uiv.p2.best-single@0.1.delta\endcsname{0}
\expandafter\def\csname pct@rev.pol.cat.uiv.p2.best-single@0.1.delta\endcsname{$-$0.2}
\expandafter\def\csname ppt@rev.pol.cat.uiv.p2.best-single@0.1.delta\endcsname{$-$0.2}
\expandafter\def\csname val@rev.pol.cat.uiv.p2.best-single@0.1.mean\endcsname{0.828}
\expandafter\def\csname cnt@rev.pol.cat.uiv.p2.best-single@0.1.mean\endcsname{1}
\expandafter\def\csname pct@rev.pol.cat.uiv.p2.best-single@0.1.mean\endcsname{82.8}
\expandafter\def\csname ppt@rev.pol.cat.uiv.p2.best-single@0.1.mean\endcsname{$+$82.8}
\expandafter\def\csname val@rev.pol.cat.uiv.p2.best-single@0.1.mean20\endcsname{0.830}
\expandafter\def\csname cnt@rev.pol.cat.uiv.p2.best-single@0.1.mean20\endcsname{1}
\expandafter\def\csname pct@rev.pol.cat.uiv.p2.best-single@0.1.mean20\endcsname{83.0}
\expandafter\def\csname ppt@rev.pol.cat.uiv.p2.best-single@0.1.mean20\endcsname{$+$83.0}
\expandafter\def\csname val@rev.pol.cat.uiv.p2.best-single@0.1.sd\endcsname{0.016}
\expandafter\def\csname cnt@rev.pol.cat.uiv.p2.best-single@0.1.sd\endcsname{0}
\expandafter\def\csname pct@rev.pol.cat.uiv.p2.best-single@0.1.sd\endcsname{1.6}
\expandafter\def\csname ppt@rev.pol.cat.uiv.p2.best-single@0.1.sd\endcsname{$+$1.6}
\expandafter\def\csname val@rev.pol.cat.uiv.p2.best-single@0.1.sd20\endcsname{0.013}
\expandafter\def\csname cnt@rev.pol.cat.uiv.p2.best-single@0.1.sd20\endcsname{0}
\expandafter\def\csname pct@rev.pol.cat.uiv.p2.best-single@0.1.sd20\endcsname{1.3}
\expandafter\def\csname ppt@rev.pol.cat.uiv.p2.best-single@0.1.sd20\endcsname{$+$1.3}
\expandafter\def\csname val@rev.pol.cat.uiv.p2.best-single@0.25.delta\endcsname{-0.005}
\expandafter\def\csname cnt@rev.pol.cat.uiv.p2.best-single@0.25.delta\endcsname{0}
\expandafter\def\csname pct@rev.pol.cat.uiv.p2.best-single@0.25.delta\endcsname{$-$0.5}
\expandafter\def\csname ppt@rev.pol.cat.uiv.p2.best-single@0.25.delta\endcsname{$-$0.5}
\expandafter\def\csname val@rev.pol.cat.uiv.p2.best-single@0.25.mean\endcsname{0.826}
\expandafter\def\csname cnt@rev.pol.cat.uiv.p2.best-single@0.25.mean\endcsname{1}
\expandafter\def\csname pct@rev.pol.cat.uiv.p2.best-single@0.25.mean\endcsname{82.6}
\expandafter\def\csname ppt@rev.pol.cat.uiv.p2.best-single@0.25.mean\endcsname{$+$82.6}
\expandafter\def\csname val@rev.pol.cat.uiv.p2.best-single@0.25.mean20\endcsname{0.831}
\expandafter\def\csname cnt@rev.pol.cat.uiv.p2.best-single@0.25.mean20\endcsname{1}
\expandafter\def\csname pct@rev.pol.cat.uiv.p2.best-single@0.25.mean20\endcsname{83.1}
\expandafter\def\csname ppt@rev.pol.cat.uiv.p2.best-single@0.25.mean20\endcsname{$+$83.1}
\expandafter\def\csname val@rev.pol.cat.uiv.p2.best-single@0.25.sd\endcsname{0.017}
\expandafter\def\csname cnt@rev.pol.cat.uiv.p2.best-single@0.25.sd\endcsname{0}
\expandafter\def\csname pct@rev.pol.cat.uiv.p2.best-single@0.25.sd\endcsname{1.7}
\expandafter\def\csname ppt@rev.pol.cat.uiv.p2.best-single@0.25.sd\endcsname{$+$1.7}
\expandafter\def\csname val@rev.pol.cat.uiv.p2.best-single@0.25.sd20\endcsname{0.012}
\expandafter\def\csname cnt@rev.pol.cat.uiv.p2.best-single@0.25.sd20\endcsname{0}
\expandafter\def\csname pct@rev.pol.cat.uiv.p2.best-single@0.25.sd20\endcsname{1.2}
\expandafter\def\csname ppt@rev.pol.cat.uiv.p2.best-single@0.25.sd20\endcsname{$+$1.2}
\expandafter\def\csname val@rev.pol.cat.uiv.p2.best-single@0.5.delta\endcsname{-0.001}
\expandafter\def\csname cnt@rev.pol.cat.uiv.p2.best-single@0.5.delta\endcsname{0}
\expandafter\def\csname pct@rev.pol.cat.uiv.p2.best-single@0.5.delta\endcsname{$-$0.1}
\expandafter\def\csname ppt@rev.pol.cat.uiv.p2.best-single@0.5.delta\endcsname{$-$0.1}
\expandafter\def\csname val@rev.pol.cat.uiv.p2.best-single@0.5.mean\endcsname{0.826}
\expandafter\def\csname cnt@rev.pol.cat.uiv.p2.best-single@0.5.mean\endcsname{1}
\expandafter\def\csname pct@rev.pol.cat.uiv.p2.best-single@0.5.mean\endcsname{82.6}
\expandafter\def\csname ppt@rev.pol.cat.uiv.p2.best-single@0.5.mean\endcsname{$+$82.6}
\expandafter\def\csname val@rev.pol.cat.uiv.p2.best-single@0.5.mean20\endcsname{0.826}
\expandafter\def\csname cnt@rev.pol.cat.uiv.p2.best-single@0.5.mean20\endcsname{1}
\expandafter\def\csname pct@rev.pol.cat.uiv.p2.best-single@0.5.mean20\endcsname{82.6}
\expandafter\def\csname ppt@rev.pol.cat.uiv.p2.best-single@0.5.mean20\endcsname{$+$82.6}
\expandafter\def\csname val@rev.pol.cat.uiv.p2.best-single@0.5.sd\endcsname{0.015}
\expandafter\def\csname cnt@rev.pol.cat.uiv.p2.best-single@0.5.sd\endcsname{0}
\expandafter\def\csname pct@rev.pol.cat.uiv.p2.best-single@0.5.sd\endcsname{1.5}
\expandafter\def\csname ppt@rev.pol.cat.uiv.p2.best-single@0.5.sd\endcsname{$+$1.5}
\expandafter\def\csname val@rev.pol.cat.uiv.p2.best-single@0.5.sd20\endcsname{0.013}
\expandafter\def\csname cnt@rev.pol.cat.uiv.p2.best-single@0.5.sd20\endcsname{0}
\expandafter\def\csname pct@rev.pol.cat.uiv.p2.best-single@0.5.sd20\endcsname{1.3}
\expandafter\def\csname ppt@rev.pol.cat.uiv.p2.best-single@0.5.sd20\endcsname{$+$1.3}
\expandafter\def\csname val@rev.pol.cat.uiv.p2.best-single@1.0.delta\endcsname{0.000}
\expandafter\def\csname cnt@rev.pol.cat.uiv.p2.best-single@1.0.delta\endcsname{0}
\expandafter\def\csname pct@rev.pol.cat.uiv.p2.best-single@1.0.delta\endcsname{0.0}
\expandafter\def\csname ppt@rev.pol.cat.uiv.p2.best-single@1.0.delta\endcsname{0.0}
\expandafter\def\csname val@rev.pol.cat.uiv.p2.best-single@1.0.mean\endcsname{0.812}
\expandafter\def\csname cnt@rev.pol.cat.uiv.p2.best-single@1.0.mean\endcsname{1}
\expandafter\def\csname pct@rev.pol.cat.uiv.p2.best-single@1.0.mean\endcsname{81.2}
\expandafter\def\csname ppt@rev.pol.cat.uiv.p2.best-single@1.0.mean\endcsname{$+$81.2}
\expandafter\def\csname val@rev.pol.cat.uiv.p2.best-single@1.0.mean20\endcsname{0.812}
\expandafter\def\csname cnt@rev.pol.cat.uiv.p2.best-single@1.0.mean20\endcsname{1}
\expandafter\def\csname pct@rev.pol.cat.uiv.p2.best-single@1.0.mean20\endcsname{81.2}
\expandafter\def\csname ppt@rev.pol.cat.uiv.p2.best-single@1.0.mean20\endcsname{$+$81.2}
\expandafter\def\csname val@rev.pol.cat.uiv.p2.best-single@1.0.sd\endcsname{0.000}
\expandafter\def\csname cnt@rev.pol.cat.uiv.p2.best-single@1.0.sd\endcsname{0}
\expandafter\def\csname pct@rev.pol.cat.uiv.p2.best-single@1.0.sd\endcsname{0.0}
\expandafter\def\csname ppt@rev.pol.cat.uiv.p2.best-single@1.0.sd\endcsname{0.0}
\expandafter\def\csname val@rev.pol.cat.uiv.p2.best-single@1.0.sd20\endcsname{0.000}
\expandafter\def\csname cnt@rev.pol.cat.uiv.p2.best-single@1.0.sd20\endcsname{0}
\expandafter\def\csname pct@rev.pol.cat.uiv.p2.best-single@1.0.sd20\endcsname{0.0}
\expandafter\def\csname ppt@rev.pol.cat.uiv.p2.best-single@1.0.sd20\endcsname{0.0}
\expandafter\def\csname val@rev.pol.cat.uiv.p2.generic-ppi-source@0.1.delta\endcsname{0.003}
\expandafter\def\csname cnt@rev.pol.cat.uiv.p2.generic-ppi-source@0.1.delta\endcsname{0}
\expandafter\def\csname pct@rev.pol.cat.uiv.p2.generic-ppi-source@0.1.delta\endcsname{0.3}
\expandafter\def\csname ppt@rev.pol.cat.uiv.p2.generic-ppi-source@0.1.delta\endcsname{$+$0.3}
\expandafter\def\csname val@rev.pol.cat.uiv.p2.generic-ppi-source@0.1.mean\endcsname{0.831}
\expandafter\def\csname cnt@rev.pol.cat.uiv.p2.generic-ppi-source@0.1.mean\endcsname{1}
\expandafter\def\csname pct@rev.pol.cat.uiv.p2.generic-ppi-source@0.1.mean\endcsname{83.1}
\expandafter\def\csname ppt@rev.pol.cat.uiv.p2.generic-ppi-source@0.1.mean\endcsname{$+$83.1}
\expandafter\def\csname val@rev.pol.cat.uiv.p2.generic-ppi-source@0.1.mean20\endcsname{0.828}
\expandafter\def\csname cnt@rev.pol.cat.uiv.p2.generic-ppi-source@0.1.mean20\endcsname{1}
\expandafter\def\csname pct@rev.pol.cat.uiv.p2.generic-ppi-source@0.1.mean20\endcsname{82.8}
\expandafter\def\csname ppt@rev.pol.cat.uiv.p2.generic-ppi-source@0.1.mean20\endcsname{$+$82.8}
\expandafter\def\csname val@rev.pol.cat.uiv.p2.generic-ppi-source@0.1.sd\endcsname{0.009}
\expandafter\def\csname cnt@rev.pol.cat.uiv.p2.generic-ppi-source@0.1.sd\endcsname{0}
\expandafter\def\csname pct@rev.pol.cat.uiv.p2.generic-ppi-source@0.1.sd\endcsname{0.9}
\expandafter\def\csname ppt@rev.pol.cat.uiv.p2.generic-ppi-source@0.1.sd\endcsname{$+$0.9}
\expandafter\def\csname val@rev.pol.cat.uiv.p2.generic-ppi-source@0.1.sd20\endcsname{0.010}
\expandafter\def\csname cnt@rev.pol.cat.uiv.p2.generic-ppi-source@0.1.sd20\endcsname{0}
\expandafter\def\csname pct@rev.pol.cat.uiv.p2.generic-ppi-source@0.1.sd20\endcsname{1.0}
\expandafter\def\csname ppt@rev.pol.cat.uiv.p2.generic-ppi-source@0.1.sd20\endcsname{$+$1.0}
\expandafter\def\csname val@rev.pol.cat.uiv.p2.generic-ppi-source@0.25.delta\endcsname{0.001}
\expandafter\def\csname cnt@rev.pol.cat.uiv.p2.generic-ppi-source@0.25.delta\endcsname{0}
\expandafter\def\csname pct@rev.pol.cat.uiv.p2.generic-ppi-source@0.25.delta\endcsname{0.1}
\expandafter\def\csname ppt@rev.pol.cat.uiv.p2.generic-ppi-source@0.25.delta\endcsname{$+$0.1}
\expandafter\def\csname val@rev.pol.cat.uiv.p2.generic-ppi-source@0.25.mean\endcsname{0.827}
\expandafter\def\csname cnt@rev.pol.cat.uiv.p2.generic-ppi-source@0.25.mean\endcsname{1}
\expandafter\def\csname pct@rev.pol.cat.uiv.p2.generic-ppi-source@0.25.mean\endcsname{82.7}
\expandafter\def\csname ppt@rev.pol.cat.uiv.p2.generic-ppi-source@0.25.mean\endcsname{$+$82.7}
\expandafter\def\csname val@rev.pol.cat.uiv.p2.generic-ppi-source@0.25.mean20\endcsname{0.826}
\expandafter\def\csname cnt@rev.pol.cat.uiv.p2.generic-ppi-source@0.25.mean20\endcsname{1}
\expandafter\def\csname pct@rev.pol.cat.uiv.p2.generic-ppi-source@0.25.mean20\endcsname{82.6}
\expandafter\def\csname ppt@rev.pol.cat.uiv.p2.generic-ppi-source@0.25.mean20\endcsname{$+$82.6}
\expandafter\def\csname val@rev.pol.cat.uiv.p2.generic-ppi-source@0.25.sd\endcsname{0.009}
\expandafter\def\csname cnt@rev.pol.cat.uiv.p2.generic-ppi-source@0.25.sd\endcsname{0}
\expandafter\def\csname pct@rev.pol.cat.uiv.p2.generic-ppi-source@0.25.sd\endcsname{0.9}
\expandafter\def\csname ppt@rev.pol.cat.uiv.p2.generic-ppi-source@0.25.sd\endcsname{$+$0.9}
\expandafter\def\csname val@rev.pol.cat.uiv.p2.generic-ppi-source@0.25.sd20\endcsname{0.009}
\expandafter\def\csname cnt@rev.pol.cat.uiv.p2.generic-ppi-source@0.25.sd20\endcsname{0}
\expandafter\def\csname pct@rev.pol.cat.uiv.p2.generic-ppi-source@0.25.sd20\endcsname{0.9}
\expandafter\def\csname ppt@rev.pol.cat.uiv.p2.generic-ppi-source@0.25.sd20\endcsname{$+$0.9}
\expandafter\def\csname val@rev.pol.cat.uiv.p2.generic-ppi-source@0.5.delta\endcsname{0.001}
\expandafter\def\csname cnt@rev.pol.cat.uiv.p2.generic-ppi-source@0.5.delta\endcsname{0}
\expandafter\def\csname pct@rev.pol.cat.uiv.p2.generic-ppi-source@0.5.delta\endcsname{0.1}
\expandafter\def\csname ppt@rev.pol.cat.uiv.p2.generic-ppi-source@0.5.delta\endcsname{$+$0.1}
\expandafter\def\csname val@rev.pol.cat.uiv.p2.generic-ppi-source@0.5.mean\endcsname{0.827}
\expandafter\def\csname cnt@rev.pol.cat.uiv.p2.generic-ppi-source@0.5.mean\endcsname{1}
\expandafter\def\csname pct@rev.pol.cat.uiv.p2.generic-ppi-source@0.5.mean\endcsname{82.7}
\expandafter\def\csname ppt@rev.pol.cat.uiv.p2.generic-ppi-source@0.5.mean\endcsname{$+$82.7}
\expandafter\def\csname val@rev.pol.cat.uiv.p2.generic-ppi-source@0.5.mean20\endcsname{0.827}
\expandafter\def\csname cnt@rev.pol.cat.uiv.p2.generic-ppi-source@0.5.mean20\endcsname{1}
\expandafter\def\csname pct@rev.pol.cat.uiv.p2.generic-ppi-source@0.5.mean20\endcsname{82.7}
\expandafter\def\csname ppt@rev.pol.cat.uiv.p2.generic-ppi-source@0.5.mean20\endcsname{$+$82.7}
\expandafter\def\csname val@rev.pol.cat.uiv.p2.generic-ppi-source@0.5.sd\endcsname{0.008}
\expandafter\def\csname cnt@rev.pol.cat.uiv.p2.generic-ppi-source@0.5.sd\endcsname{0}
\expandafter\def\csname pct@rev.pol.cat.uiv.p2.generic-ppi-source@0.5.sd\endcsname{0.8}
\expandafter\def\csname ppt@rev.pol.cat.uiv.p2.generic-ppi-source@0.5.sd\endcsname{$+$0.8}
\expandafter\def\csname val@rev.pol.cat.uiv.p2.generic-ppi-source@0.5.sd20\endcsname{0.008}
\expandafter\def\csname cnt@rev.pol.cat.uiv.p2.generic-ppi-source@0.5.sd20\endcsname{0}
\expandafter\def\csname pct@rev.pol.cat.uiv.p2.generic-ppi-source@0.5.sd20\endcsname{0.8}
\expandafter\def\csname ppt@rev.pol.cat.uiv.p2.generic-ppi-source@0.5.sd20\endcsname{$+$0.8}
\expandafter\def\csname val@rev.pol.cat.uiv.p2.generic-ppi-source@1.0.delta\endcsname{0.000}
\expandafter\def\csname cnt@rev.pol.cat.uiv.p2.generic-ppi-source@1.0.delta\endcsname{0}
\expandafter\def\csname pct@rev.pol.cat.uiv.p2.generic-ppi-source@1.0.delta\endcsname{0.0}
\expandafter\def\csname ppt@rev.pol.cat.uiv.p2.generic-ppi-source@1.0.delta\endcsname{0.0}
\expandafter\def\csname val@rev.pol.cat.uiv.p2.generic-ppi-source@1.0.mean\endcsname{0.832}
\expandafter\def\csname cnt@rev.pol.cat.uiv.p2.generic-ppi-source@1.0.mean\endcsname{1}
\expandafter\def\csname pct@rev.pol.cat.uiv.p2.generic-ppi-source@1.0.mean\endcsname{83.2}
\expandafter\def\csname ppt@rev.pol.cat.uiv.p2.generic-ppi-source@1.0.mean\endcsname{$+$83.2}
\expandafter\def\csname val@rev.pol.cat.uiv.p2.generic-ppi-source@1.0.mean20\endcsname{0.832}
\expandafter\def\csname cnt@rev.pol.cat.uiv.p2.generic-ppi-source@1.0.mean20\endcsname{1}
\expandafter\def\csname pct@rev.pol.cat.uiv.p2.generic-ppi-source@1.0.mean20\endcsname{83.2}
\expandafter\def\csname ppt@rev.pol.cat.uiv.p2.generic-ppi-source@1.0.mean20\endcsname{$+$83.2}
\expandafter\def\csname val@rev.pol.cat.uiv.p2.generic-ppi-source@1.0.sd\endcsname{0.000}
\expandafter\def\csname cnt@rev.pol.cat.uiv.p2.generic-ppi-source@1.0.sd\endcsname{0}
\expandafter\def\csname pct@rev.pol.cat.uiv.p2.generic-ppi-source@1.0.sd\endcsname{0.0}
\expandafter\def\csname ppt@rev.pol.cat.uiv.p2.generic-ppi-source@1.0.sd\endcsname{0.0}
\expandafter\def\csname val@rev.pol.cat.uiv.p2.generic-ppi-source@1.0.sd20\endcsname{0.000}
\expandafter\def\csname cnt@rev.pol.cat.uiv.p2.generic-ppi-source@1.0.sd20\endcsname{0}
\expandafter\def\csname pct@rev.pol.cat.uiv.p2.generic-ppi-source@1.0.sd20\endcsname{0.0}
\expandafter\def\csname ppt@rev.pol.cat.uiv.p2.generic-ppi-source@1.0.sd20\endcsname{0.0}
\expandafter\def\csname val@rev.pol.cat.uiv.p2.rcr-stratified@0.1.delta\endcsname{0.001}
\expandafter\def\csname cnt@rev.pol.cat.uiv.p2.rcr-stratified@0.1.delta\endcsname{0}
\expandafter\def\csname pct@rev.pol.cat.uiv.p2.rcr-stratified@0.1.delta\endcsname{0.1}
\expandafter\def\csname ppt@rev.pol.cat.uiv.p2.rcr-stratified@0.1.delta\endcsname{$+$0.1}
\expandafter\def\csname val@rev.pol.cat.uiv.p2.rcr-stratified@0.1.mean\endcsname{0.826}
\expandafter\def\csname cnt@rev.pol.cat.uiv.p2.rcr-stratified@0.1.mean\endcsname{1}
\expandafter\def\csname pct@rev.pol.cat.uiv.p2.rcr-stratified@0.1.mean\endcsname{82.6}
\expandafter\def\csname ppt@rev.pol.cat.uiv.p2.rcr-stratified@0.1.mean\endcsname{$+$82.6}
\expandafter\def\csname val@rev.pol.cat.uiv.p2.rcr-stratified@0.1.mean20\endcsname{0.826}
\expandafter\def\csname cnt@rev.pol.cat.uiv.p2.rcr-stratified@0.1.mean20\endcsname{1}
\expandafter\def\csname pct@rev.pol.cat.uiv.p2.rcr-stratified@0.1.mean20\endcsname{82.6}
\expandafter\def\csname ppt@rev.pol.cat.uiv.p2.rcr-stratified@0.1.mean20\endcsname{$+$82.6}
\expandafter\def\csname val@rev.pol.cat.uiv.p2.rcr-stratified@0.1.sd\endcsname{0.008}
\expandafter\def\csname cnt@rev.pol.cat.uiv.p2.rcr-stratified@0.1.sd\endcsname{0}
\expandafter\def\csname pct@rev.pol.cat.uiv.p2.rcr-stratified@0.1.sd\endcsname{0.8}
\expandafter\def\csname ppt@rev.pol.cat.uiv.p2.rcr-stratified@0.1.sd\endcsname{$+$0.8}
\expandafter\def\csname val@rev.pol.cat.uiv.p2.rcr-stratified@0.1.sd20\endcsname{0.009}
\expandafter\def\csname cnt@rev.pol.cat.uiv.p2.rcr-stratified@0.1.sd20\endcsname{0}
\expandafter\def\csname pct@rev.pol.cat.uiv.p2.rcr-stratified@0.1.sd20\endcsname{0.9}
\expandafter\def\csname ppt@rev.pol.cat.uiv.p2.rcr-stratified@0.1.sd20\endcsname{$+$0.9}
\expandafter\def\csname val@rev.pol.cat.uiv.p2.rcr-stratified@0.25.delta\endcsname{-0.001}
\expandafter\def\csname cnt@rev.pol.cat.uiv.p2.rcr-stratified@0.25.delta\endcsname{0}
\expandafter\def\csname pct@rev.pol.cat.uiv.p2.rcr-stratified@0.25.delta\endcsname{$-$0.1}
\expandafter\def\csname ppt@rev.pol.cat.uiv.p2.rcr-stratified@0.25.delta\endcsname{$-$0.1}
\expandafter\def\csname val@rev.pol.cat.uiv.p2.rcr-stratified@0.25.mean\endcsname{0.828}
\expandafter\def\csname cnt@rev.pol.cat.uiv.p2.rcr-stratified@0.25.mean\endcsname{1}
\expandafter\def\csname pct@rev.pol.cat.uiv.p2.rcr-stratified@0.25.mean\endcsname{82.8}
\expandafter\def\csname ppt@rev.pol.cat.uiv.p2.rcr-stratified@0.25.mean\endcsname{$+$82.8}
\expandafter\def\csname val@rev.pol.cat.uiv.p2.rcr-stratified@0.25.mean20\endcsname{0.829}
\expandafter\def\csname cnt@rev.pol.cat.uiv.p2.rcr-stratified@0.25.mean20\endcsname{1}
\expandafter\def\csname pct@rev.pol.cat.uiv.p2.rcr-stratified@0.25.mean20\endcsname{82.9}
\expandafter\def\csname ppt@rev.pol.cat.uiv.p2.rcr-stratified@0.25.mean20\endcsname{$+$82.9}
\expandafter\def\csname val@rev.pol.cat.uiv.p2.rcr-stratified@0.25.sd\endcsname{0.008}
\expandafter\def\csname cnt@rev.pol.cat.uiv.p2.rcr-stratified@0.25.sd\endcsname{0}
\expandafter\def\csname pct@rev.pol.cat.uiv.p2.rcr-stratified@0.25.sd\endcsname{0.8}
\expandafter\def\csname ppt@rev.pol.cat.uiv.p2.rcr-stratified@0.25.sd\endcsname{$+$0.8}
\expandafter\def\csname val@rev.pol.cat.uiv.p2.rcr-stratified@0.25.sd20\endcsname{0.009}
\expandafter\def\csname cnt@rev.pol.cat.uiv.p2.rcr-stratified@0.25.sd20\endcsname{0}
\expandafter\def\csname pct@rev.pol.cat.uiv.p2.rcr-stratified@0.25.sd20\endcsname{0.9}
\expandafter\def\csname ppt@rev.pol.cat.uiv.p2.rcr-stratified@0.25.sd20\endcsname{$+$0.9}
\expandafter\def\csname val@rev.pol.cat.uiv.p2.rcr-stratified@0.5.delta\endcsname{-0.004}
\expandafter\def\csname cnt@rev.pol.cat.uiv.p2.rcr-stratified@0.5.delta\endcsname{0}
\expandafter\def\csname pct@rev.pol.cat.uiv.p2.rcr-stratified@0.5.delta\endcsname{$-$0.4}
\expandafter\def\csname ppt@rev.pol.cat.uiv.p2.rcr-stratified@0.5.delta\endcsname{$-$0.4}
\expandafter\def\csname val@rev.pol.cat.uiv.p2.rcr-stratified@0.5.mean\endcsname{0.829}
\expandafter\def\csname cnt@rev.pol.cat.uiv.p2.rcr-stratified@0.5.mean\endcsname{1}
\expandafter\def\csname pct@rev.pol.cat.uiv.p2.rcr-stratified@0.5.mean\endcsname{82.9}
\expandafter\def\csname ppt@rev.pol.cat.uiv.p2.rcr-stratified@0.5.mean\endcsname{$+$82.9}
\expandafter\def\csname val@rev.pol.cat.uiv.p2.rcr-stratified@0.5.mean20\endcsname{0.832}
\expandafter\def\csname cnt@rev.pol.cat.uiv.p2.rcr-stratified@0.5.mean20\endcsname{1}
\expandafter\def\csname pct@rev.pol.cat.uiv.p2.rcr-stratified@0.5.mean20\endcsname{83.2}
\expandafter\def\csname ppt@rev.pol.cat.uiv.p2.rcr-stratified@0.5.mean20\endcsname{$+$83.2}
\expandafter\def\csname val@rev.pol.cat.uiv.p2.rcr-stratified@0.5.sd\endcsname{0.008}
\expandafter\def\csname cnt@rev.pol.cat.uiv.p2.rcr-stratified@0.5.sd\endcsname{0}
\expandafter\def\csname pct@rev.pol.cat.uiv.p2.rcr-stratified@0.5.sd\endcsname{0.8}
\expandafter\def\csname ppt@rev.pol.cat.uiv.p2.rcr-stratified@0.5.sd\endcsname{$+$0.8}
\expandafter\def\csname val@rev.pol.cat.uiv.p2.rcr-stratified@0.5.sd20\endcsname{0.007}
\expandafter\def\csname cnt@rev.pol.cat.uiv.p2.rcr-stratified@0.5.sd20\endcsname{0}
\expandafter\def\csname pct@rev.pol.cat.uiv.p2.rcr-stratified@0.5.sd20\endcsname{0.7}
\expandafter\def\csname ppt@rev.pol.cat.uiv.p2.rcr-stratified@0.5.sd20\endcsname{$+$0.7}
\expandafter\def\csname val@rev.pol.cat.uiv.p2.rcr-stratified@1.0.delta\endcsname{0.000}
\expandafter\def\csname cnt@rev.pol.cat.uiv.p2.rcr-stratified@1.0.delta\endcsname{0}
\expandafter\def\csname pct@rev.pol.cat.uiv.p2.rcr-stratified@1.0.delta\endcsname{0.0}
\expandafter\def\csname ppt@rev.pol.cat.uiv.p2.rcr-stratified@1.0.delta\endcsname{0.0}
\expandafter\def\csname val@rev.pol.cat.uiv.p2.rcr-stratified@1.0.mean\endcsname{0.832}
\expandafter\def\csname cnt@rev.pol.cat.uiv.p2.rcr-stratified@1.0.mean\endcsname{1}
\expandafter\def\csname pct@rev.pol.cat.uiv.p2.rcr-stratified@1.0.mean\endcsname{83.2}
\expandafter\def\csname ppt@rev.pol.cat.uiv.p2.rcr-stratified@1.0.mean\endcsname{$+$83.2}
\expandafter\def\csname val@rev.pol.cat.uiv.p2.rcr-stratified@1.0.mean20\endcsname{0.832}
\expandafter\def\csname cnt@rev.pol.cat.uiv.p2.rcr-stratified@1.0.mean20\endcsname{1}
\expandafter\def\csname pct@rev.pol.cat.uiv.p2.rcr-stratified@1.0.mean20\endcsname{83.2}
\expandafter\def\csname ppt@rev.pol.cat.uiv.p2.rcr-stratified@1.0.mean20\endcsname{$+$83.2}
\expandafter\def\csname val@rev.pol.cat.uiv.p2.rcr-stratified@1.0.sd\endcsname{0.000}
\expandafter\def\csname cnt@rev.pol.cat.uiv.p2.rcr-stratified@1.0.sd\endcsname{0}
\expandafter\def\csname pct@rev.pol.cat.uiv.p2.rcr-stratified@1.0.sd\endcsname{0.0}
\expandafter\def\csname ppt@rev.pol.cat.uiv.p2.rcr-stratified@1.0.sd\endcsname{0.0}
\expandafter\def\csname val@rev.pol.cat.uiv.p2.rcr-stratified@1.0.sd20\endcsname{0.000}
\expandafter\def\csname cnt@rev.pol.cat.uiv.p2.rcr-stratified@1.0.sd20\endcsname{0}
\expandafter\def\csname pct@rev.pol.cat.uiv.p2.rcr-stratified@1.0.sd20\endcsname{0.0}
\expandafter\def\csname ppt@rev.pol.cat.uiv.p2.rcr-stratified@1.0.sd20\endcsname{0.0}
\expandafter\def\csname val@rev.pol.cat.uiv.p2.replay-ppi-uniform@0.1.delta\endcsname{0.002}
\expandafter\def\csname cnt@rev.pol.cat.uiv.p2.replay-ppi-uniform@0.1.delta\endcsname{0}
\expandafter\def\csname pct@rev.pol.cat.uiv.p2.replay-ppi-uniform@0.1.delta\endcsname{0.2}
\expandafter\def\csname ppt@rev.pol.cat.uiv.p2.replay-ppi-uniform@0.1.delta\endcsname{$+$0.2}
\expandafter\def\csname val@rev.pol.cat.uiv.p2.replay-ppi-uniform@0.1.mean\endcsname{0.829}
\expandafter\def\csname cnt@rev.pol.cat.uiv.p2.replay-ppi-uniform@0.1.mean\endcsname{1}
\expandafter\def\csname pct@rev.pol.cat.uiv.p2.replay-ppi-uniform@0.1.mean\endcsname{82.9}
\expandafter\def\csname ppt@rev.pol.cat.uiv.p2.replay-ppi-uniform@0.1.mean\endcsname{$+$82.9}
\expandafter\def\csname val@rev.pol.cat.uiv.p2.replay-ppi-uniform@0.1.mean20\endcsname{0.826}
\expandafter\def\csname cnt@rev.pol.cat.uiv.p2.replay-ppi-uniform@0.1.mean20\endcsname{1}
\expandafter\def\csname pct@rev.pol.cat.uiv.p2.replay-ppi-uniform@0.1.mean20\endcsname{82.6}
\expandafter\def\csname ppt@rev.pol.cat.uiv.p2.replay-ppi-uniform@0.1.mean20\endcsname{$+$82.6}
\expandafter\def\csname val@rev.pol.cat.uiv.p2.replay-ppi-uniform@0.1.sd\endcsname{0.008}
\expandafter\def\csname cnt@rev.pol.cat.uiv.p2.replay-ppi-uniform@0.1.sd\endcsname{0}
\expandafter\def\csname pct@rev.pol.cat.uiv.p2.replay-ppi-uniform@0.1.sd\endcsname{0.8}
\expandafter\def\csname ppt@rev.pol.cat.uiv.p2.replay-ppi-uniform@0.1.sd\endcsname{$+$0.8}
\expandafter\def\csname val@rev.pol.cat.uiv.p2.replay-ppi-uniform@0.1.sd20\endcsname{0.009}
\expandafter\def\csname cnt@rev.pol.cat.uiv.p2.replay-ppi-uniform@0.1.sd20\endcsname{0}
\expandafter\def\csname pct@rev.pol.cat.uiv.p2.replay-ppi-uniform@0.1.sd20\endcsname{0.9}
\expandafter\def\csname ppt@rev.pol.cat.uiv.p2.replay-ppi-uniform@0.1.sd20\endcsname{$+$0.9}
\expandafter\def\csname val@rev.pol.cat.uiv.p2.replay-ppi-uniform@0.25.delta\endcsname{-0.000}
\expandafter\def\csname cnt@rev.pol.cat.uiv.p2.replay-ppi-uniform@0.25.delta\endcsname{0}
\expandafter\def\csname pct@rev.pol.cat.uiv.p2.replay-ppi-uniform@0.25.delta\endcsname{0.0}
\expandafter\def\csname ppt@rev.pol.cat.uiv.p2.replay-ppi-uniform@0.25.delta\endcsname{0.0}
\expandafter\def\csname val@rev.pol.cat.uiv.p2.replay-ppi-uniform@0.25.mean\endcsname{0.827}
\expandafter\def\csname cnt@rev.pol.cat.uiv.p2.replay-ppi-uniform@0.25.mean\endcsname{1}
\expandafter\def\csname pct@rev.pol.cat.uiv.p2.replay-ppi-uniform@0.25.mean\endcsname{82.7}
\expandafter\def\csname ppt@rev.pol.cat.uiv.p2.replay-ppi-uniform@0.25.mean\endcsname{$+$82.7}
\expandafter\def\csname val@rev.pol.cat.uiv.p2.replay-ppi-uniform@0.25.mean20\endcsname{0.828}
\expandafter\def\csname cnt@rev.pol.cat.uiv.p2.replay-ppi-uniform@0.25.mean20\endcsname{1}
\expandafter\def\csname pct@rev.pol.cat.uiv.p2.replay-ppi-uniform@0.25.mean20\endcsname{82.8}
\expandafter\def\csname ppt@rev.pol.cat.uiv.p2.replay-ppi-uniform@0.25.mean20\endcsname{$+$82.8}
\expandafter\def\csname val@rev.pol.cat.uiv.p2.replay-ppi-uniform@0.25.sd\endcsname{0.009}
\expandafter\def\csname cnt@rev.pol.cat.uiv.p2.replay-ppi-uniform@0.25.sd\endcsname{0}
\expandafter\def\csname pct@rev.pol.cat.uiv.p2.replay-ppi-uniform@0.25.sd\endcsname{0.9}
\expandafter\def\csname ppt@rev.pol.cat.uiv.p2.replay-ppi-uniform@0.25.sd\endcsname{$+$0.9}
\expandafter\def\csname val@rev.pol.cat.uiv.p2.replay-ppi-uniform@0.25.sd20\endcsname{0.008}
\expandafter\def\csname cnt@rev.pol.cat.uiv.p2.replay-ppi-uniform@0.25.sd20\endcsname{0}
\expandafter\def\csname pct@rev.pol.cat.uiv.p2.replay-ppi-uniform@0.25.sd20\endcsname{0.8}
\expandafter\def\csname ppt@rev.pol.cat.uiv.p2.replay-ppi-uniform@0.25.sd20\endcsname{$+$0.8}
\expandafter\def\csname val@rev.pol.cat.uiv.p2.replay-ppi-uniform@0.5.delta\endcsname{0.000}
\expandafter\def\csname cnt@rev.pol.cat.uiv.p2.replay-ppi-uniform@0.5.delta\endcsname{0}
\expandafter\def\csname pct@rev.pol.cat.uiv.p2.replay-ppi-uniform@0.5.delta\endcsname{0.0}
\expandafter\def\csname ppt@rev.pol.cat.uiv.p2.replay-ppi-uniform@0.5.delta\endcsname{0.0}
\expandafter\def\csname val@rev.pol.cat.uiv.p2.replay-ppi-uniform@0.5.mean\endcsname{0.826}
\expandafter\def\csname cnt@rev.pol.cat.uiv.p2.replay-ppi-uniform@0.5.mean\endcsname{1}
\expandafter\def\csname pct@rev.pol.cat.uiv.p2.replay-ppi-uniform@0.5.mean\endcsname{82.6}
\expandafter\def\csname ppt@rev.pol.cat.uiv.p2.replay-ppi-uniform@0.5.mean\endcsname{$+$82.6}
\expandafter\def\csname val@rev.pol.cat.uiv.p2.replay-ppi-uniform@0.5.mean20\endcsname{0.826}
\expandafter\def\csname cnt@rev.pol.cat.uiv.p2.replay-ppi-uniform@0.5.mean20\endcsname{1}
\expandafter\def\csname pct@rev.pol.cat.uiv.p2.replay-ppi-uniform@0.5.mean20\endcsname{82.6}
\expandafter\def\csname ppt@rev.pol.cat.uiv.p2.replay-ppi-uniform@0.5.mean20\endcsname{$+$82.6}
\expandafter\def\csname val@rev.pol.cat.uiv.p2.replay-ppi-uniform@0.5.sd\endcsname{0.009}
\expandafter\def\csname cnt@rev.pol.cat.uiv.p2.replay-ppi-uniform@0.5.sd\endcsname{0}
\expandafter\def\csname pct@rev.pol.cat.uiv.p2.replay-ppi-uniform@0.5.sd\endcsname{0.9}
\expandafter\def\csname ppt@rev.pol.cat.uiv.p2.replay-ppi-uniform@0.5.sd\endcsname{$+$0.9}
\expandafter\def\csname val@rev.pol.cat.uiv.p2.replay-ppi-uniform@0.5.sd20\endcsname{0.008}
\expandafter\def\csname cnt@rev.pol.cat.uiv.p2.replay-ppi-uniform@0.5.sd20\endcsname{0}
\expandafter\def\csname pct@rev.pol.cat.uiv.p2.replay-ppi-uniform@0.5.sd20\endcsname{0.8}
\expandafter\def\csname ppt@rev.pol.cat.uiv.p2.replay-ppi-uniform@0.5.sd20\endcsname{$+$0.8}
\expandafter\def\csname val@rev.pol.cat.uiv.p2.replay-ppi-uniform@1.0.delta\endcsname{0.000}
\expandafter\def\csname cnt@rev.pol.cat.uiv.p2.replay-ppi-uniform@1.0.delta\endcsname{0}
\expandafter\def\csname pct@rev.pol.cat.uiv.p2.replay-ppi-uniform@1.0.delta\endcsname{0.0}
\expandafter\def\csname ppt@rev.pol.cat.uiv.p2.replay-ppi-uniform@1.0.delta\endcsname{0.0}
\expandafter\def\csname val@rev.pol.cat.uiv.p2.replay-ppi-uniform@1.0.mean\endcsname{0.832}
\expandafter\def\csname cnt@rev.pol.cat.uiv.p2.replay-ppi-uniform@1.0.mean\endcsname{1}
\expandafter\def\csname pct@rev.pol.cat.uiv.p2.replay-ppi-uniform@1.0.mean\endcsname{83.2}
\expandafter\def\csname ppt@rev.pol.cat.uiv.p2.replay-ppi-uniform@1.0.mean\endcsname{$+$83.2}
\expandafter\def\csname val@rev.pol.cat.uiv.p2.replay-ppi-uniform@1.0.mean20\endcsname{0.832}
\expandafter\def\csname cnt@rev.pol.cat.uiv.p2.replay-ppi-uniform@1.0.mean20\endcsname{1}
\expandafter\def\csname pct@rev.pol.cat.uiv.p2.replay-ppi-uniform@1.0.mean20\endcsname{83.2}
\expandafter\def\csname ppt@rev.pol.cat.uiv.p2.replay-ppi-uniform@1.0.mean20\endcsname{$+$83.2}
\expandafter\def\csname val@rev.pol.cat.uiv.p2.replay-ppi-uniform@1.0.sd\endcsname{0.000}
\expandafter\def\csname cnt@rev.pol.cat.uiv.p2.replay-ppi-uniform@1.0.sd\endcsname{0}
\expandafter\def\csname pct@rev.pol.cat.uiv.p2.replay-ppi-uniform@1.0.sd\endcsname{0.0}
\expandafter\def\csname ppt@rev.pol.cat.uiv.p2.replay-ppi-uniform@1.0.sd\endcsname{0.0}
\expandafter\def\csname val@rev.pol.cat.uiv.p2.replay-ppi-uniform@1.0.sd20\endcsname{0.000}
\expandafter\def\csname cnt@rev.pol.cat.uiv.p2.replay-ppi-uniform@1.0.sd20\endcsname{0}
\expandafter\def\csname pct@rev.pol.cat.uiv.p2.replay-ppi-uniform@1.0.sd20\endcsname{0.0}
\expandafter\def\csname ppt@rev.pol.cat.uiv.p2.replay-ppi-uniform@1.0.sd20\endcsname{0.0}
\expandafter\def\csname val@rev.pol.cat.uiv.p2.uniform-fresh@0.1.delta\endcsname{0.003}
\expandafter\def\csname cnt@rev.pol.cat.uiv.p2.uniform-fresh@0.1.delta\endcsname{0}
\expandafter\def\csname pct@rev.pol.cat.uiv.p2.uniform-fresh@0.1.delta\endcsname{0.3}
\expandafter\def\csname ppt@rev.pol.cat.uiv.p2.uniform-fresh@0.1.delta\endcsname{$+$0.3}
\expandafter\def\csname val@rev.pol.cat.uiv.p2.uniform-fresh@0.1.mean\endcsname{0.822}
\expandafter\def\csname cnt@rev.pol.cat.uiv.p2.uniform-fresh@0.1.mean\endcsname{1}
\expandafter\def\csname pct@rev.pol.cat.uiv.p2.uniform-fresh@0.1.mean\endcsname{82.2}
\expandafter\def\csname ppt@rev.pol.cat.uiv.p2.uniform-fresh@0.1.mean\endcsname{$+$82.2}
\expandafter\def\csname val@rev.pol.cat.uiv.p2.uniform-fresh@0.1.mean20\endcsname{0.819}
\expandafter\def\csname cnt@rev.pol.cat.uiv.p2.uniform-fresh@0.1.mean20\endcsname{1}
\expandafter\def\csname pct@rev.pol.cat.uiv.p2.uniform-fresh@0.1.mean20\endcsname{81.9}
\expandafter\def\csname ppt@rev.pol.cat.uiv.p2.uniform-fresh@0.1.mean20\endcsname{$+$81.9}
\expandafter\def\csname val@rev.pol.cat.uiv.p2.uniform-fresh@0.1.sd\endcsname{0.009}
\expandafter\def\csname cnt@rev.pol.cat.uiv.p2.uniform-fresh@0.1.sd\endcsname{0}
\expandafter\def\csname pct@rev.pol.cat.uiv.p2.uniform-fresh@0.1.sd\endcsname{0.9}
\expandafter\def\csname ppt@rev.pol.cat.uiv.p2.uniform-fresh@0.1.sd\endcsname{$+$0.9}
\expandafter\def\csname val@rev.pol.cat.uiv.p2.uniform-fresh@0.1.sd20\endcsname{0.007}
\expandafter\def\csname cnt@rev.pol.cat.uiv.p2.uniform-fresh@0.1.sd20\endcsname{0}
\expandafter\def\csname pct@rev.pol.cat.uiv.p2.uniform-fresh@0.1.sd20\endcsname{0.7}
\expandafter\def\csname ppt@rev.pol.cat.uiv.p2.uniform-fresh@0.1.sd20\endcsname{$+$0.7}
\expandafter\def\csname val@rev.pol.cat.uiv.p2.uniform-fresh@0.25.delta\endcsname{0.003}
\expandafter\def\csname cnt@rev.pol.cat.uiv.p2.uniform-fresh@0.25.delta\endcsname{0}
\expandafter\def\csname pct@rev.pol.cat.uiv.p2.uniform-fresh@0.25.delta\endcsname{0.3}
\expandafter\def\csname ppt@rev.pol.cat.uiv.p2.uniform-fresh@0.25.delta\endcsname{$+$0.3}
\expandafter\def\csname val@rev.pol.cat.uiv.p2.uniform-fresh@0.25.mean\endcsname{0.824}
\expandafter\def\csname cnt@rev.pol.cat.uiv.p2.uniform-fresh@0.25.mean\endcsname{1}
\expandafter\def\csname pct@rev.pol.cat.uiv.p2.uniform-fresh@0.25.mean\endcsname{82.4}
\expandafter\def\csname ppt@rev.pol.cat.uiv.p2.uniform-fresh@0.25.mean\endcsname{$+$82.4}
\expandafter\def\csname val@rev.pol.cat.uiv.p2.uniform-fresh@0.25.mean20\endcsname{0.821}
\expandafter\def\csname cnt@rev.pol.cat.uiv.p2.uniform-fresh@0.25.mean20\endcsname{1}
\expandafter\def\csname pct@rev.pol.cat.uiv.p2.uniform-fresh@0.25.mean20\endcsname{82.1}
\expandafter\def\csname ppt@rev.pol.cat.uiv.p2.uniform-fresh@0.25.mean20\endcsname{$+$82.1}
\expandafter\def\csname val@rev.pol.cat.uiv.p2.uniform-fresh@0.25.sd\endcsname{0.009}
\expandafter\def\csname cnt@rev.pol.cat.uiv.p2.uniform-fresh@0.25.sd\endcsname{0}
\expandafter\def\csname pct@rev.pol.cat.uiv.p2.uniform-fresh@0.25.sd\endcsname{0.9}
\expandafter\def\csname ppt@rev.pol.cat.uiv.p2.uniform-fresh@0.25.sd\endcsname{$+$0.9}
\expandafter\def\csname val@rev.pol.cat.uiv.p2.uniform-fresh@0.25.sd20\endcsname{0.009}
\expandafter\def\csname cnt@rev.pol.cat.uiv.p2.uniform-fresh@0.25.sd20\endcsname{0}
\expandafter\def\csname pct@rev.pol.cat.uiv.p2.uniform-fresh@0.25.sd20\endcsname{0.9}
\expandafter\def\csname ppt@rev.pol.cat.uiv.p2.uniform-fresh@0.25.sd20\endcsname{$+$0.9}
\expandafter\def\csname val@rev.pol.cat.uiv.p2.uniform-fresh@0.5.delta\endcsname{-0.004}
\expandafter\def\csname cnt@rev.pol.cat.uiv.p2.uniform-fresh@0.5.delta\endcsname{0}
\expandafter\def\csname pct@rev.pol.cat.uiv.p2.uniform-fresh@0.5.delta\endcsname{$-$0.4}
\expandafter\def\csname ppt@rev.pol.cat.uiv.p2.uniform-fresh@0.5.delta\endcsname{$-$0.4}
\expandafter\def\csname val@rev.pol.cat.uiv.p2.uniform-fresh@0.5.mean\endcsname{0.825}
\expandafter\def\csname cnt@rev.pol.cat.uiv.p2.uniform-fresh@0.5.mean\endcsname{1}
\expandafter\def\csname pct@rev.pol.cat.uiv.p2.uniform-fresh@0.5.mean\endcsname{82.5}
\expandafter\def\csname ppt@rev.pol.cat.uiv.p2.uniform-fresh@0.5.mean\endcsname{$+$82.5}
\expandafter\def\csname val@rev.pol.cat.uiv.p2.uniform-fresh@0.5.mean20\endcsname{0.829}
\expandafter\def\csname cnt@rev.pol.cat.uiv.p2.uniform-fresh@0.5.mean20\endcsname{1}
\expandafter\def\csname pct@rev.pol.cat.uiv.p2.uniform-fresh@0.5.mean20\endcsname{82.9}
\expandafter\def\csname ppt@rev.pol.cat.uiv.p2.uniform-fresh@0.5.mean20\endcsname{$+$82.9}
\expandafter\def\csname val@rev.pol.cat.uiv.p2.uniform-fresh@0.5.sd\endcsname{0.009}
\expandafter\def\csname cnt@rev.pol.cat.uiv.p2.uniform-fresh@0.5.sd\endcsname{0}
\expandafter\def\csname pct@rev.pol.cat.uiv.p2.uniform-fresh@0.5.sd\endcsname{0.9}
\expandafter\def\csname ppt@rev.pol.cat.uiv.p2.uniform-fresh@0.5.sd\endcsname{$+$0.9}
\expandafter\def\csname val@rev.pol.cat.uiv.p2.uniform-fresh@0.5.sd20\endcsname{0.008}
\expandafter\def\csname cnt@rev.pol.cat.uiv.p2.uniform-fresh@0.5.sd20\endcsname{0}
\expandafter\def\csname pct@rev.pol.cat.uiv.p2.uniform-fresh@0.5.sd20\endcsname{0.8}
\expandafter\def\csname ppt@rev.pol.cat.uiv.p2.uniform-fresh@0.5.sd20\endcsname{$+$0.8}
\expandafter\def\csname val@rev.pol.cat.uiv.p2.uniform-fresh@1.0.delta\endcsname{0.000}
\expandafter\def\csname cnt@rev.pol.cat.uiv.p2.uniform-fresh@1.0.delta\endcsname{0}
\expandafter\def\csname pct@rev.pol.cat.uiv.p2.uniform-fresh@1.0.delta\endcsname{0.0}
\expandafter\def\csname ppt@rev.pol.cat.uiv.p2.uniform-fresh@1.0.delta\endcsname{0.0}
\expandafter\def\csname val@rev.pol.cat.uiv.p2.uniform-fresh@1.0.mean\endcsname{0.832}
\expandafter\def\csname cnt@rev.pol.cat.uiv.p2.uniform-fresh@1.0.mean\endcsname{1}
\expandafter\def\csname pct@rev.pol.cat.uiv.p2.uniform-fresh@1.0.mean\endcsname{83.2}
\expandafter\def\csname ppt@rev.pol.cat.uiv.p2.uniform-fresh@1.0.mean\endcsname{$+$83.2}
\expandafter\def\csname val@rev.pol.cat.uiv.p2.uniform-fresh@1.0.mean20\endcsname{0.832}
\expandafter\def\csname cnt@rev.pol.cat.uiv.p2.uniform-fresh@1.0.mean20\endcsname{1}
\expandafter\def\csname pct@rev.pol.cat.uiv.p2.uniform-fresh@1.0.mean20\endcsname{83.2}
\expandafter\def\csname ppt@rev.pol.cat.uiv.p2.uniform-fresh@1.0.mean20\endcsname{$+$83.2}
\expandafter\def\csname val@rev.pol.cat.uiv.p2.uniform-fresh@1.0.sd\endcsname{0.000}
\expandafter\def\csname cnt@rev.pol.cat.uiv.p2.uniform-fresh@1.0.sd\endcsname{0}
\expandafter\def\csname pct@rev.pol.cat.uiv.p2.uniform-fresh@1.0.sd\endcsname{0.0}
\expandafter\def\csname ppt@rev.pol.cat.uiv.p2.uniform-fresh@1.0.sd\endcsname{0.0}
\expandafter\def\csname val@rev.pol.cat.uiv.p2.uniform-fresh@1.0.sd20\endcsname{0.000}
\expandafter\def\csname cnt@rev.pol.cat.uiv.p2.uniform-fresh@1.0.sd20\endcsname{0}
\expandafter\def\csname pct@rev.pol.cat.uiv.p2.uniform-fresh@1.0.sd20\endcsname{0.0}
\expandafter\def\csname ppt@rev.pol.cat.uiv.p2.uniform-fresh@1.0.sd20\endcsname{0.0}
\expandafter\def\csname val@rev.pol.eqc.best-single.max\endcsname{0.016}
\expandafter\def\csname cnt@rev.pol.eqc.best-single.max\endcsname{0}
\expandafter\def\csname pct@rev.pol.eqc.best-single.max\endcsname{1.6}
\expandafter\def\csname ppt@rev.pol.eqc.best-single.max\endcsname{$+$1.6}
\expandafter\def\csname val@rev.pol.eqc.best-single.min\endcsname{-0.018}
\expandafter\def\csname cnt@rev.pol.eqc.best-single.min\endcsname{0}
\expandafter\def\csname pct@rev.pol.eqc.best-single.min\endcsname{$-$1.8}
\expandafter\def\csname ppt@rev.pol.eqc.best-single.min\endcsname{$-$1.8}
\expandafter\def\csname val@rev.pol.eqc.best-single.n-ci-above-0\endcsname{1}
\expandafter\def\csname cnt@rev.pol.eqc.best-single.n-ci-above-0\endcsname{1}
\expandafter\def\csname pct@rev.pol.eqc.best-single.n-ci-above-0\endcsname{100.0}
\expandafter\def\csname ppt@rev.pol.eqc.best-single.n-ci-above-0\endcsname{$+$100.0}
\expandafter\def\csname val@rev.pol.eqc.best-single.n-ci-above-0-lin20\endcsname{0}
\expandafter\def\csname cnt@rev.pol.eqc.best-single.n-ci-above-0-lin20\endcsname{0}
\expandafter\def\csname pct@rev.pol.eqc.best-single.n-ci-above-0-lin20\endcsname{0.0}
\expandafter\def\csname ppt@rev.pol.eqc.best-single.n-ci-above-0-lin20\endcsname{0.0}
\expandafter\def\csname val@rev.pol.eqc.best-single.n-ci-above-0-linN\endcsname{0}
\expandafter\def\csname cnt@rev.pol.eqc.best-single.n-ci-above-0-linN\endcsname{0}
\expandafter\def\csname pct@rev.pol.eqc.best-single.n-ci-above-0-linN\endcsname{0.0}
\expandafter\def\csname ppt@rev.pol.eqc.best-single.n-ci-above-0-linN\endcsname{0.0}
\expandafter\def\csname val@rev.pol.eqc.best-single.n-ci-below-0\endcsname{1}
\expandafter\def\csname cnt@rev.pol.eqc.best-single.n-ci-below-0\endcsname{1}
\expandafter\def\csname pct@rev.pol.eqc.best-single.n-ci-below-0\endcsname{100.0}
\expandafter\def\csname ppt@rev.pol.eqc.best-single.n-ci-below-0\endcsname{$+$100.0}
\expandafter\def\csname val@rev.pol.eqc.best-single.n-ci-below-0-lin20\endcsname{2}
\expandafter\def\csname cnt@rev.pol.eqc.best-single.n-ci-below-0-lin20\endcsname{2}
\expandafter\def\csname pct@rev.pol.eqc.best-single.n-ci-below-0-lin20\endcsname{200.0}
\expandafter\def\csname ppt@rev.pol.eqc.best-single.n-ci-below-0-lin20\endcsname{$+$200.0}
\expandafter\def\csname val@rev.pol.eqc.best-single.n-ci-below-0-linN\endcsname{1}
\expandafter\def\csname cnt@rev.pol.eqc.best-single.n-ci-below-0-linN\endcsname{1}
\expandafter\def\csname pct@rev.pol.eqc.best-single.n-ci-below-0-linN\endcsname{100.0}
\expandafter\def\csname ppt@rev.pol.eqc.best-single.n-ci-below-0-linN\endcsname{$+$100.0}
\expandafter\def\csname val@rev.pol.eqc.best-single.n-sign-change-vs-lin\endcsname{1}
\expandafter\def\csname cnt@rev.pol.eqc.best-single.n-sign-change-vs-lin\endcsname{1}
\expandafter\def\csname pct@rev.pol.eqc.best-single.n-sign-change-vs-lin\endcsname{100.0}
\expandafter\def\csname ppt@rev.pol.eqc.best-single.n-sign-change-vs-lin\endcsname{$+$100.0}
\expandafter\def\csname val@rev.pol.eqc.generic-ppi-source.max\endcsname{0.002}
\expandafter\def\csname cnt@rev.pol.eqc.generic-ppi-source.max\endcsname{0}
\expandafter\def\csname pct@rev.pol.eqc.generic-ppi-source.max\endcsname{0.2}
\expandafter\def\csname ppt@rev.pol.eqc.generic-ppi-source.max\endcsname{$+$0.2}
\expandafter\def\csname val@rev.pol.eqc.generic-ppi-source.min\endcsname{-0.011}
\expandafter\def\csname cnt@rev.pol.eqc.generic-ppi-source.min\endcsname{0}
\expandafter\def\csname pct@rev.pol.eqc.generic-ppi-source.min\endcsname{$-$1.1}
\expandafter\def\csname ppt@rev.pol.eqc.generic-ppi-source.min\endcsname{$-$1.1}
\expandafter\def\csname val@rev.pol.eqc.generic-ppi-source.n-ci-above-0\endcsname{0}
\expandafter\def\csname cnt@rev.pol.eqc.generic-ppi-source.n-ci-above-0\endcsname{0}
\expandafter\def\csname pct@rev.pol.eqc.generic-ppi-source.n-ci-above-0\endcsname{0.0}
\expandafter\def\csname ppt@rev.pol.eqc.generic-ppi-source.n-ci-above-0\endcsname{0.0}
\expandafter\def\csname val@rev.pol.eqc.generic-ppi-source.n-ci-above-0-lin20\endcsname{1}
\expandafter\def\csname cnt@rev.pol.eqc.generic-ppi-source.n-ci-above-0-lin20\endcsname{1}
\expandafter\def\csname pct@rev.pol.eqc.generic-ppi-source.n-ci-above-0-lin20\endcsname{100.0}
\expandafter\def\csname ppt@rev.pol.eqc.generic-ppi-source.n-ci-above-0-lin20\endcsname{$+$100.0}
\expandafter\def\csname val@rev.pol.eqc.generic-ppi-source.n-ci-above-0-linN\endcsname{1}
\expandafter\def\csname cnt@rev.pol.eqc.generic-ppi-source.n-ci-above-0-linN\endcsname{1}
\expandafter\def\csname pct@rev.pol.eqc.generic-ppi-source.n-ci-above-0-linN\endcsname{100.0}
\expandafter\def\csname ppt@rev.pol.eqc.generic-ppi-source.n-ci-above-0-linN\endcsname{$+$100.0}
\expandafter\def\csname val@rev.pol.eqc.generic-ppi-source.n-ci-below-0\endcsname{2}
\expandafter\def\csname cnt@rev.pol.eqc.generic-ppi-source.n-ci-below-0\endcsname{2}
\expandafter\def\csname pct@rev.pol.eqc.generic-ppi-source.n-ci-below-0\endcsname{200.0}
\expandafter\def\csname ppt@rev.pol.eqc.generic-ppi-source.n-ci-below-0\endcsname{$+$200.0}
\expandafter\def\csname val@rev.pol.eqc.generic-ppi-source.n-ci-below-0-lin20\endcsname{2}
\expandafter\def\csname cnt@rev.pol.eqc.generic-ppi-source.n-ci-below-0-lin20\endcsname{2}
\expandafter\def\csname pct@rev.pol.eqc.generic-ppi-source.n-ci-below-0-lin20\endcsname{200.0}
\expandafter\def\csname ppt@rev.pol.eqc.generic-ppi-source.n-ci-below-0-lin20\endcsname{$+$200.0}
\expandafter\def\csname val@rev.pol.eqc.generic-ppi-source.n-ci-below-0-linN\endcsname{2}
\expandafter\def\csname cnt@rev.pol.eqc.generic-ppi-source.n-ci-below-0-linN\endcsname{2}
\expandafter\def\csname pct@rev.pol.eqc.generic-ppi-source.n-ci-below-0-linN\endcsname{200.0}
\expandafter\def\csname ppt@rev.pol.eqc.generic-ppi-source.n-ci-below-0-linN\endcsname{$+$200.0}
\expandafter\def\csname val@rev.pol.eqc.generic-ppi-source.n-sign-change-vs-lin\endcsname{0}
\expandafter\def\csname cnt@rev.pol.eqc.generic-ppi-source.n-sign-change-vs-lin\endcsname{0}
\expandafter\def\csname pct@rev.pol.eqc.generic-ppi-source.n-sign-change-vs-lin\endcsname{0.0}
\expandafter\def\csname ppt@rev.pol.eqc.generic-ppi-source.n-sign-change-vs-lin\endcsname{0.0}
\expandafter\def\csname val@rev.pol.eqc.replay-ppi-uniform.max\endcsname{0.004}
\expandafter\def\csname cnt@rev.pol.eqc.replay-ppi-uniform.max\endcsname{0}
\expandafter\def\csname pct@rev.pol.eqc.replay-ppi-uniform.max\endcsname{0.4}
\expandafter\def\csname ppt@rev.pol.eqc.replay-ppi-uniform.max\endcsname{$+$0.4}
\expandafter\def\csname val@rev.pol.eqc.replay-ppi-uniform.min\endcsname{0.000}
\expandafter\def\csname cnt@rev.pol.eqc.replay-ppi-uniform.min\endcsname{0}
\expandafter\def\csname pct@rev.pol.eqc.replay-ppi-uniform.min\endcsname{0.0}
\expandafter\def\csname ppt@rev.pol.eqc.replay-ppi-uniform.min\endcsname{0.0}
\expandafter\def\csname val@rev.pol.eqc.replay-ppi-uniform.n-ci-above-0\endcsname{3}
\expandafter\def\csname cnt@rev.pol.eqc.replay-ppi-uniform.n-ci-above-0\endcsname{3}
\expandafter\def\csname pct@rev.pol.eqc.replay-ppi-uniform.n-ci-above-0\endcsname{300.0}
\expandafter\def\csname ppt@rev.pol.eqc.replay-ppi-uniform.n-ci-above-0\endcsname{$+$300.0}
\expandafter\def\csname val@rev.pol.eqc.replay-ppi-uniform.n-ci-above-0-lin20\endcsname{3}
\expandafter\def\csname cnt@rev.pol.eqc.replay-ppi-uniform.n-ci-above-0-lin20\endcsname{3}
\expandafter\def\csname pct@rev.pol.eqc.replay-ppi-uniform.n-ci-above-0-lin20\endcsname{300.0}
\expandafter\def\csname ppt@rev.pol.eqc.replay-ppi-uniform.n-ci-above-0-lin20\endcsname{$+$300.0}
\expandafter\def\csname val@rev.pol.eqc.replay-ppi-uniform.n-ci-above-0-linN\endcsname{3}
\expandafter\def\csname cnt@rev.pol.eqc.replay-ppi-uniform.n-ci-above-0-linN\endcsname{3}
\expandafter\def\csname pct@rev.pol.eqc.replay-ppi-uniform.n-ci-above-0-linN\endcsname{300.0}
\expandafter\def\csname ppt@rev.pol.eqc.replay-ppi-uniform.n-ci-above-0-linN\endcsname{$+$300.0}
\expandafter\def\csname val@rev.pol.eqc.replay-ppi-uniform.n-ci-below-0\endcsname{0}
\expandafter\def\csname cnt@rev.pol.eqc.replay-ppi-uniform.n-ci-below-0\endcsname{0}
\expandafter\def\csname pct@rev.pol.eqc.replay-ppi-uniform.n-ci-below-0\endcsname{0.0}
\expandafter\def\csname ppt@rev.pol.eqc.replay-ppi-uniform.n-ci-below-0\endcsname{0.0}
\expandafter\def\csname val@rev.pol.eqc.replay-ppi-uniform.n-ci-below-0-lin20\endcsname{0}
\expandafter\def\csname cnt@rev.pol.eqc.replay-ppi-uniform.n-ci-below-0-lin20\endcsname{0}
\expandafter\def\csname pct@rev.pol.eqc.replay-ppi-uniform.n-ci-below-0-lin20\endcsname{0.0}
\expandafter\def\csname ppt@rev.pol.eqc.replay-ppi-uniform.n-ci-below-0-lin20\endcsname{0.0}
\expandafter\def\csname val@rev.pol.eqc.replay-ppi-uniform.n-ci-below-0-linN\endcsname{0}
\expandafter\def\csname cnt@rev.pol.eqc.replay-ppi-uniform.n-ci-below-0-linN\endcsname{0}
\expandafter\def\csname pct@rev.pol.eqc.replay-ppi-uniform.n-ci-below-0-linN\endcsname{0.0}
\expandafter\def\csname ppt@rev.pol.eqc.replay-ppi-uniform.n-ci-below-0-linN\endcsname{0.0}
\expandafter\def\csname val@rev.pol.eqc.replay-ppi-uniform.n-sign-change-vs-lin\endcsname{0}
\expandafter\def\csname cnt@rev.pol.eqc.replay-ppi-uniform.n-sign-change-vs-lin\endcsname{0}
\expandafter\def\csname pct@rev.pol.eqc.replay-ppi-uniform.n-sign-change-vs-lin\endcsname{0.0}
\expandafter\def\csname ppt@rev.pol.eqc.replay-ppi-uniform.n-sign-change-vs-lin\endcsname{0.0}
\expandafter\def\csname val@rev.pol.eqc.sspro.p1.best-single.cost\endcsname{128.912}
\expandafter\def\csname cnt@rev.pol.eqc.sspro.p1.best-single.cost\endcsname{129}
\expandafter\def\csname pct@rev.pol.eqc.sspro.p1.best-single.cost\endcsname{12891.2}
\expandafter\def\csname ppt@rev.pol.eqc.sspro.p1.best-single.cost\endcsname{$+$12891.2}
\expandafter\def\csname val@rev.pol.eqc.sspro.p1.best-single.lin.ci95.0\endcsname{-0.022}
\expandafter\def\csname cnt@rev.pol.eqc.sspro.p1.best-single.lin.ci95.0\endcsname{0}
\expandafter\def\csname pct@rev.pol.eqc.sspro.p1.best-single.lin.ci95.0\endcsname{$-$2.2}
\expandafter\def\csname ppt@rev.pol.eqc.sspro.p1.best-single.lin.ci95.0\endcsname{$-$2.2}
\expandafter\def\csname val@rev.pol.eqc.sspro.p1.best-single.lin.ci95.1\endcsname{0.045}
\expandafter\def\csname cnt@rev.pol.eqc.sspro.p1.best-single.lin.ci95.1\endcsname{0}
\expandafter\def\csname pct@rev.pol.eqc.sspro.p1.best-single.lin.ci95.1\endcsname{4.5}
\expandafter\def\csname ppt@rev.pol.eqc.sspro.p1.best-single.lin.ci95.1\endcsname{$+$4.5}
\expandafter\def\csname val@rev.pol.eqc.sspro.p1.best-single.lin.mean\endcsname{0.009}
\expandafter\def\csname cnt@rev.pol.eqc.sspro.p1.best-single.lin.mean\endcsname{0}
\expandafter\def\csname pct@rev.pol.eqc.sspro.p1.best-single.lin.mean\endcsname{0.9}
\expandafter\def\csname ppt@rev.pol.eqc.sspro.p1.best-single.lin.mean\endcsname{$+$0.9}
\expandafter\def\csname val@rev.pol.eqc.sspro.p1.best-single.linN.ci95.0\endcsname{-0.023}
\expandafter\def\csname cnt@rev.pol.eqc.sspro.p1.best-single.linN.ci95.0\endcsname{0}
\expandafter\def\csname pct@rev.pol.eqc.sspro.p1.best-single.linN.ci95.0\endcsname{$-$2.3}
\expandafter\def\csname ppt@rev.pol.eqc.sspro.p1.best-single.linN.ci95.0\endcsname{$-$2.3}
\expandafter\def\csname val@rev.pol.eqc.sspro.p1.best-single.linN.ci95.1\endcsname{0.048}
\expandafter\def\csname cnt@rev.pol.eqc.sspro.p1.best-single.linN.ci95.1\endcsname{0}
\expandafter\def\csname pct@rev.pol.eqc.sspro.p1.best-single.linN.ci95.1\endcsname{4.8}
\expandafter\def\csname ppt@rev.pol.eqc.sspro.p1.best-single.linN.ci95.1\endcsname{$+$4.8}
\expandafter\def\csname val@rev.pol.eqc.sspro.p1.best-single.linN.mean\endcsname{0.010}
\expandafter\def\csname cnt@rev.pol.eqc.sspro.p1.best-single.linN.mean\endcsname{0}
\expandafter\def\csname pct@rev.pol.eqc.sspro.p1.best-single.linN.mean\endcsname{1.0}
\expandafter\def\csname ppt@rev.pol.eqc.sspro.p1.best-single.linN.mean\endcsname{$+$1.0}
\expandafter\def\csname val@rev.pol.eqc.sspro.p1.best-single.minus.ci95.0\endcsname{-0.016}
\expandafter\def\csname cnt@rev.pol.eqc.sspro.p1.best-single.minus.ci95.0\endcsname{0}
\expandafter\def\csname pct@rev.pol.eqc.sspro.p1.best-single.minus.ci95.0\endcsname{$-$1.6}
\expandafter\def\csname ppt@rev.pol.eqc.sspro.p1.best-single.minus.ci95.0\endcsname{$-$1.6}
\expandafter\def\csname val@rev.pol.eqc.sspro.p1.best-single.minus.ci95.1\endcsname{0.023}
\expandafter\def\csname cnt@rev.pol.eqc.sspro.p1.best-single.minus.ci95.1\endcsname{0}
\expandafter\def\csname pct@rev.pol.eqc.sspro.p1.best-single.minus.ci95.1\endcsname{2.3}
\expandafter\def\csname ppt@rev.pol.eqc.sspro.p1.best-single.minus.ci95.1\endcsname{$+$2.3}
\expandafter\def\csname val@rev.pol.eqc.sspro.p1.best-single.minus.mean\endcsname{0.004}
\expandafter\def\csname cnt@rev.pol.eqc.sspro.p1.best-single.minus.mean\endcsname{0}
\expandafter\def\csname pct@rev.pol.eqc.sspro.p1.best-single.minus.mean\endcsname{0.4}
\expandafter\def\csname ppt@rev.pol.eqc.sspro.p1.best-single.minus.mean\endcsname{$+$0.4}
\expandafter\def\csname val@rev.pol.eqc.sspro.p1.best-single.minus.p-gt\endcsname{0.332}
\expandafter\def\csname cnt@rev.pol.eqc.sspro.p1.best-single.minus.p-gt\endcsname{0}
\expandafter\def\csname pct@rev.pol.eqc.sspro.p1.best-single.minus.p-gt\endcsname{33.2}
\expandafter\def\csname ppt@rev.pol.eqc.sspro.p1.best-single.minus.p-gt\endcsname{$+$33.2}
\expandafter\def\csname val@rev.pol.eqc.sspro.p1.best-single.p-signflip-gt\endcsname{0.343}
\expandafter\def\csname cnt@rev.pol.eqc.sspro.p1.best-single.p-signflip-gt\endcsname{0}
\expandafter\def\csname pct@rev.pol.eqc.sspro.p1.best-single.p-signflip-gt\endcsname{34.3}
\expandafter\def\csname ppt@rev.pol.eqc.sspro.p1.best-single.p-signflip-gt\endcsname{$+$34.3}
\expandafter\def\csname val@rev.pol.eqc.sspro.p1.best-single.p-signflip-lt\endcsname{0.658}
\expandafter\def\csname cnt@rev.pol.eqc.sspro.p1.best-single.p-signflip-lt\endcsname{1}
\expandafter\def\csname pct@rev.pol.eqc.sspro.p1.best-single.p-signflip-lt\endcsname{65.8}
\expandafter\def\csname ppt@rev.pol.eqc.sspro.p1.best-single.p-signflip-lt\endcsname{$+$65.8}
\expandafter\def\csname val@rev.pol.eqc.sspro.p1.best-single.score\endcsname{0.851}
\expandafter\def\csname cnt@rev.pol.eqc.sspro.p1.best-single.score\endcsname{1}
\expandafter\def\csname pct@rev.pol.eqc.sspro.p1.best-single.score\endcsname{85.1}
\expandafter\def\csname ppt@rev.pol.eqc.sspro.p1.best-single.score\endcsname{$+$85.1}
\expandafter\def\csname val@rev.pol.eqc.sspro.p1.generic-ppi-source.cost\endcsname{128.912}
\expandafter\def\csname cnt@rev.pol.eqc.sspro.p1.generic-ppi-source.cost\endcsname{129}
\expandafter\def\csname pct@rev.pol.eqc.sspro.p1.generic-ppi-source.cost\endcsname{12891.2}
\expandafter\def\csname ppt@rev.pol.eqc.sspro.p1.generic-ppi-source.cost\endcsname{$+$12891.2}
\expandafter\def\csname val@rev.pol.eqc.sspro.p1.generic-ppi-source.lin.ci95.0\endcsname{-0.004}
\expandafter\def\csname cnt@rev.pol.eqc.sspro.p1.generic-ppi-source.lin.ci95.0\endcsname{0}
\expandafter\def\csname pct@rev.pol.eqc.sspro.p1.generic-ppi-source.lin.ci95.0\endcsname{$-$0.4}
\expandafter\def\csname ppt@rev.pol.eqc.sspro.p1.generic-ppi-source.lin.ci95.0\endcsname{$-$0.4}
\expandafter\def\csname val@rev.pol.eqc.sspro.p1.generic-ppi-source.lin.ci95.1\endcsname{-0.000}
\expandafter\def\csname cnt@rev.pol.eqc.sspro.p1.generic-ppi-source.lin.ci95.1\endcsname{0}
\expandafter\def\csname pct@rev.pol.eqc.sspro.p1.generic-ppi-source.lin.ci95.1\endcsname{0.0}
\expandafter\def\csname ppt@rev.pol.eqc.sspro.p1.generic-ppi-source.lin.ci95.1\endcsname{0.0}
\expandafter\def\csname val@rev.pol.eqc.sspro.p1.generic-ppi-source.lin.mean\endcsname{-0.002}
\expandafter\def\csname cnt@rev.pol.eqc.sspro.p1.generic-ppi-source.lin.mean\endcsname{0}
\expandafter\def\csname pct@rev.pol.eqc.sspro.p1.generic-ppi-source.lin.mean\endcsname{$-$0.2}
\expandafter\def\csname ppt@rev.pol.eqc.sspro.p1.generic-ppi-source.lin.mean\endcsname{$-$0.2}
\expandafter\def\csname val@rev.pol.eqc.sspro.p1.generic-ppi-source.linN.ci95.0\endcsname{-0.006}
\expandafter\def\csname cnt@rev.pol.eqc.sspro.p1.generic-ppi-source.linN.ci95.0\endcsname{0}
\expandafter\def\csname pct@rev.pol.eqc.sspro.p1.generic-ppi-source.linN.ci95.0\endcsname{$-$0.6}
\expandafter\def\csname ppt@rev.pol.eqc.sspro.p1.generic-ppi-source.linN.ci95.0\endcsname{$-$0.6}
\expandafter\def\csname val@rev.pol.eqc.sspro.p1.generic-ppi-source.linN.ci95.1\endcsname{-0.000}
\expandafter\def\csname cnt@rev.pol.eqc.sspro.p1.generic-ppi-source.linN.ci95.1\endcsname{0}
\expandafter\def\csname pct@rev.pol.eqc.sspro.p1.generic-ppi-source.linN.ci95.1\endcsname{0.0}
\expandafter\def\csname ppt@rev.pol.eqc.sspro.p1.generic-ppi-source.linN.ci95.1\endcsname{0.0}
\expandafter\def\csname val@rev.pol.eqc.sspro.p1.generic-ppi-source.linN.mean\endcsname{-0.003}
\expandafter\def\csname cnt@rev.pol.eqc.sspro.p1.generic-ppi-source.linN.mean\endcsname{0}
\expandafter\def\csname pct@rev.pol.eqc.sspro.p1.generic-ppi-source.linN.mean\endcsname{$-$0.3}
\expandafter\def\csname ppt@rev.pol.eqc.sspro.p1.generic-ppi-source.linN.mean\endcsname{$-$0.3}
\expandafter\def\csname val@rev.pol.eqc.sspro.p1.generic-ppi-source.minus.ci95.0\endcsname{-0.008}
\expandafter\def\csname cnt@rev.pol.eqc.sspro.p1.generic-ppi-source.minus.ci95.0\endcsname{0}
\expandafter\def\csname pct@rev.pol.eqc.sspro.p1.generic-ppi-source.minus.ci95.0\endcsname{$-$0.8}
\expandafter\def\csname ppt@rev.pol.eqc.sspro.p1.generic-ppi-source.minus.ci95.0\endcsname{$-$0.8}
\expandafter\def\csname val@rev.pol.eqc.sspro.p1.generic-ppi-source.minus.ci95.1\endcsname{-0.000}
\expandafter\def\csname cnt@rev.pol.eqc.sspro.p1.generic-ppi-source.minus.ci95.1\endcsname{0}
\expandafter\def\csname pct@rev.pol.eqc.sspro.p1.generic-ppi-source.minus.ci95.1\endcsname{0.0}
\expandafter\def\csname ppt@rev.pol.eqc.sspro.p1.generic-ppi-source.minus.ci95.1\endcsname{0.0}
\expandafter\def\csname val@rev.pol.eqc.sspro.p1.generic-ppi-source.minus.mean\endcsname{-0.004}
\expandafter\def\csname cnt@rev.pol.eqc.sspro.p1.generic-ppi-source.minus.mean\endcsname{0}
\expandafter\def\csname pct@rev.pol.eqc.sspro.p1.generic-ppi-source.minus.mean\endcsname{$-$0.4}
\expandafter\def\csname ppt@rev.pol.eqc.sspro.p1.generic-ppi-source.minus.mean\endcsname{$-$0.4}
\expandafter\def\csname val@rev.pol.eqc.sspro.p1.generic-ppi-source.minus.p-gt\endcsname{0.990}
\expandafter\def\csname cnt@rev.pol.eqc.sspro.p1.generic-ppi-source.minus.p-gt\endcsname{1}
\expandafter\def\csname pct@rev.pol.eqc.sspro.p1.generic-ppi-source.minus.p-gt\endcsname{99.0}
\expandafter\def\csname ppt@rev.pol.eqc.sspro.p1.generic-ppi-source.minus.p-gt\endcsname{$+$99.0}
\expandafter\def\csname val@rev.pol.eqc.sspro.p1.generic-ppi-source.p-signflip-gt\endcsname{0.962}
\expandafter\def\csname cnt@rev.pol.eqc.sspro.p1.generic-ppi-source.p-signflip-gt\endcsname{1}
\expandafter\def\csname pct@rev.pol.eqc.sspro.p1.generic-ppi-source.p-signflip-gt\endcsname{96.2}
\expandafter\def\csname ppt@rev.pol.eqc.sspro.p1.generic-ppi-source.p-signflip-gt\endcsname{$+$96.2}
\expandafter\def\csname val@rev.pol.eqc.sspro.p1.generic-ppi-source.p-signflip-lt\endcsname{0.046}
\expandafter\def\csname cnt@rev.pol.eqc.sspro.p1.generic-ppi-source.p-signflip-lt\endcsname{0}
\expandafter\def\csname pct@rev.pol.eqc.sspro.p1.generic-ppi-source.p-signflip-lt\endcsname{4.6}
\expandafter\def\csname ppt@rev.pol.eqc.sspro.p1.generic-ppi-source.p-signflip-lt\endcsname{$+$4.6}
\expandafter\def\csname val@rev.pol.eqc.sspro.p1.generic-ppi-source.score\endcsname{0.859}
\expandafter\def\csname cnt@rev.pol.eqc.sspro.p1.generic-ppi-source.score\endcsname{1}
\expandafter\def\csname pct@rev.pol.eqc.sspro.p1.generic-ppi-source.score\endcsname{85.9}
\expandafter\def\csname ppt@rev.pol.eqc.sspro.p1.generic-ppi-source.score\endcsname{$+$85.9}
\expandafter\def\csname val@rev.pol.eqc.sspro.p1.rcr-cost\endcsname{128.911}
\expandafter\def\csname cnt@rev.pol.eqc.sspro.p1.rcr-cost\endcsname{129}
\expandafter\def\csname pct@rev.pol.eqc.sspro.p1.rcr-cost\endcsname{12891.1}
\expandafter\def\csname ppt@rev.pol.eqc.sspro.p1.rcr-cost\endcsname{$+$12891.1}
\expandafter\def\csname val@rev.pol.eqc.sspro.p1.replay-ppi-uniform.cost\endcsname{128.912}
\expandafter\def\csname cnt@rev.pol.eqc.sspro.p1.replay-ppi-uniform.cost\endcsname{129}
\expandafter\def\csname pct@rev.pol.eqc.sspro.p1.replay-ppi-uniform.cost\endcsname{12891.2}
\expandafter\def\csname ppt@rev.pol.eqc.sspro.p1.replay-ppi-uniform.cost\endcsname{$+$12891.2}
\expandafter\def\csname val@rev.pol.eqc.sspro.p1.replay-ppi-uniform.lin.ci95.0\endcsname{0.001}
\expandafter\def\csname cnt@rev.pol.eqc.sspro.p1.replay-ppi-uniform.lin.ci95.0\endcsname{0}
\expandafter\def\csname pct@rev.pol.eqc.sspro.p1.replay-ppi-uniform.lin.ci95.0\endcsname{0.1}
\expandafter\def\csname ppt@rev.pol.eqc.sspro.p1.replay-ppi-uniform.lin.ci95.0\endcsname{$+$0.1}
\expandafter\def\csname val@rev.pol.eqc.sspro.p1.replay-ppi-uniform.lin.ci95.1\endcsname{0.018}
\expandafter\def\csname cnt@rev.pol.eqc.sspro.p1.replay-ppi-uniform.lin.ci95.1\endcsname{0}
\expandafter\def\csname pct@rev.pol.eqc.sspro.p1.replay-ppi-uniform.lin.ci95.1\endcsname{1.8}
\expandafter\def\csname ppt@rev.pol.eqc.sspro.p1.replay-ppi-uniform.lin.ci95.1\endcsname{$+$1.8}
\expandafter\def\csname val@rev.pol.eqc.sspro.p1.replay-ppi-uniform.lin.mean\endcsname{0.009}
\expandafter\def\csname cnt@rev.pol.eqc.sspro.p1.replay-ppi-uniform.lin.mean\endcsname{0}
\expandafter\def\csname pct@rev.pol.eqc.sspro.p1.replay-ppi-uniform.lin.mean\endcsname{0.9}
\expandafter\def\csname ppt@rev.pol.eqc.sspro.p1.replay-ppi-uniform.lin.mean\endcsname{$+$0.9}
\expandafter\def\csname val@rev.pol.eqc.sspro.p1.replay-ppi-uniform.linN.ci95.0\endcsname{0.000}
\expandafter\def\csname cnt@rev.pol.eqc.sspro.p1.replay-ppi-uniform.linN.ci95.0\endcsname{0}
\expandafter\def\csname pct@rev.pol.eqc.sspro.p1.replay-ppi-uniform.linN.ci95.0\endcsname{0.0}
\expandafter\def\csname ppt@rev.pol.eqc.sspro.p1.replay-ppi-uniform.linN.ci95.0\endcsname{0.0}
\expandafter\def\csname val@rev.pol.eqc.sspro.p1.replay-ppi-uniform.linN.ci95.1\endcsname{0.007}
\expandafter\def\csname cnt@rev.pol.eqc.sspro.p1.replay-ppi-uniform.linN.ci95.1\endcsname{0}
\expandafter\def\csname pct@rev.pol.eqc.sspro.p1.replay-ppi-uniform.linN.ci95.1\endcsname{0.7}
\expandafter\def\csname ppt@rev.pol.eqc.sspro.p1.replay-ppi-uniform.linN.ci95.1\endcsname{$+$0.7}
\expandafter\def\csname val@rev.pol.eqc.sspro.p1.replay-ppi-uniform.linN.mean\endcsname{0.003}
\expandafter\def\csname cnt@rev.pol.eqc.sspro.p1.replay-ppi-uniform.linN.mean\endcsname{0}
\expandafter\def\csname pct@rev.pol.eqc.sspro.p1.replay-ppi-uniform.linN.mean\endcsname{0.3}
\expandafter\def\csname ppt@rev.pol.eqc.sspro.p1.replay-ppi-uniform.linN.mean\endcsname{$+$0.3}
\expandafter\def\csname val@rev.pol.eqc.sspro.p1.replay-ppi-uniform.minus.ci95.0\endcsname{0.000}
\expandafter\def\csname cnt@rev.pol.eqc.sspro.p1.replay-ppi-uniform.minus.ci95.0\endcsname{0}
\expandafter\def\csname pct@rev.pol.eqc.sspro.p1.replay-ppi-uniform.minus.ci95.0\endcsname{0.0}
\expandafter\def\csname ppt@rev.pol.eqc.sspro.p1.replay-ppi-uniform.minus.ci95.0\endcsname{0.0}
\expandafter\def\csname val@rev.pol.eqc.sspro.p1.replay-ppi-uniform.minus.ci95.1\endcsname{0.007}
\expandafter\def\csname cnt@rev.pol.eqc.sspro.p1.replay-ppi-uniform.minus.ci95.1\endcsname{0}
\expandafter\def\csname pct@rev.pol.eqc.sspro.p1.replay-ppi-uniform.minus.ci95.1\endcsname{0.7}
\expandafter\def\csname ppt@rev.pol.eqc.sspro.p1.replay-ppi-uniform.minus.ci95.1\endcsname{$+$0.7}
\expandafter\def\csname val@rev.pol.eqc.sspro.p1.replay-ppi-uniform.minus.mean\endcsname{0.003}
\expandafter\def\csname cnt@rev.pol.eqc.sspro.p1.replay-ppi-uniform.minus.mean\endcsname{0}
\expandafter\def\csname pct@rev.pol.eqc.sspro.p1.replay-ppi-uniform.minus.mean\endcsname{0.3}
\expandafter\def\csname ppt@rev.pol.eqc.sspro.p1.replay-ppi-uniform.minus.mean\endcsname{$+$0.3}
\expandafter\def\csname val@rev.pol.eqc.sspro.p1.replay-ppi-uniform.minus.p-gt\endcsname{0.009}
\expandafter\def\csname cnt@rev.pol.eqc.sspro.p1.replay-ppi-uniform.minus.p-gt\endcsname{0}
\expandafter\def\csname pct@rev.pol.eqc.sspro.p1.replay-ppi-uniform.minus.p-gt\endcsname{0.9}
\expandafter\def\csname ppt@rev.pol.eqc.sspro.p1.replay-ppi-uniform.minus.p-gt\endcsname{$+$0.9}
\expandafter\def\csname val@rev.pol.eqc.sspro.p1.replay-ppi-uniform.p-signflip-gt\endcsname{0.032}
\expandafter\def\csname cnt@rev.pol.eqc.sspro.p1.replay-ppi-uniform.p-signflip-gt\endcsname{0}
\expandafter\def\csname pct@rev.pol.eqc.sspro.p1.replay-ppi-uniform.p-signflip-gt\endcsname{3.2}
\expandafter\def\csname ppt@rev.pol.eqc.sspro.p1.replay-ppi-uniform.p-signflip-gt\endcsname{$+$3.2}
\expandafter\def\csname val@rev.pol.eqc.sspro.p1.replay-ppi-uniform.p-signflip-lt\endcsname{0.971}
\expandafter\def\csname cnt@rev.pol.eqc.sspro.p1.replay-ppi-uniform.p-signflip-lt\endcsname{1}
\expandafter\def\csname pct@rev.pol.eqc.sspro.p1.replay-ppi-uniform.p-signflip-lt\endcsname{97.1}
\expandafter\def\csname ppt@rev.pol.eqc.sspro.p1.replay-ppi-uniform.p-signflip-lt\endcsname{$+$97.1}
\expandafter\def\csname val@rev.pol.eqc.sspro.p1.replay-ppi-uniform.score\endcsname{0.852}
\expandafter\def\csname cnt@rev.pol.eqc.sspro.p1.replay-ppi-uniform.score\endcsname{1}
\expandafter\def\csname pct@rev.pol.eqc.sspro.p1.replay-ppi-uniform.score\endcsname{85.2}
\expandafter\def\csname ppt@rev.pol.eqc.sspro.p1.replay-ppi-uniform.score\endcsname{$+$85.2}
\expandafter\def\csname val@rev.pol.eqc.sspro.p1.uniform-fresh.cost\endcsname{128.912}
\expandafter\def\csname cnt@rev.pol.eqc.sspro.p1.uniform-fresh.cost\endcsname{129}
\expandafter\def\csname pct@rev.pol.eqc.sspro.p1.uniform-fresh.cost\endcsname{12891.2}
\expandafter\def\csname ppt@rev.pol.eqc.sspro.p1.uniform-fresh.cost\endcsname{$+$12891.2}
\expandafter\def\csname val@rev.pol.eqc.sspro.p1.uniform-fresh.lin.ci95.0\endcsname{-0.004}
\expandafter\def\csname cnt@rev.pol.eqc.sspro.p1.uniform-fresh.lin.ci95.0\endcsname{0}
\expandafter\def\csname pct@rev.pol.eqc.sspro.p1.uniform-fresh.lin.ci95.0\endcsname{$-$0.4}
\expandafter\def\csname ppt@rev.pol.eqc.sspro.p1.uniform-fresh.lin.ci95.0\endcsname{$-$0.4}
\expandafter\def\csname val@rev.pol.eqc.sspro.p1.uniform-fresh.lin.ci95.1\endcsname{-0.000}
\expandafter\def\csname cnt@rev.pol.eqc.sspro.p1.uniform-fresh.lin.ci95.1\endcsname{0}
\expandafter\def\csname pct@rev.pol.eqc.sspro.p1.uniform-fresh.lin.ci95.1\endcsname{0.0}
\expandafter\def\csname ppt@rev.pol.eqc.sspro.p1.uniform-fresh.lin.ci95.1\endcsname{0.0}
\expandafter\def\csname val@rev.pol.eqc.sspro.p1.uniform-fresh.lin.mean\endcsname{-0.002}
\expandafter\def\csname cnt@rev.pol.eqc.sspro.p1.uniform-fresh.lin.mean\endcsname{0}
\expandafter\def\csname pct@rev.pol.eqc.sspro.p1.uniform-fresh.lin.mean\endcsname{$-$0.2}
\expandafter\def\csname ppt@rev.pol.eqc.sspro.p1.uniform-fresh.lin.mean\endcsname{$-$0.2}
\expandafter\def\csname val@rev.pol.eqc.sspro.p1.uniform-fresh.linN.ci95.0\endcsname{-0.006}
\expandafter\def\csname cnt@rev.pol.eqc.sspro.p1.uniform-fresh.linN.ci95.0\endcsname{0}
\expandafter\def\csname pct@rev.pol.eqc.sspro.p1.uniform-fresh.linN.ci95.0\endcsname{$-$0.6}
\expandafter\def\csname ppt@rev.pol.eqc.sspro.p1.uniform-fresh.linN.ci95.0\endcsname{$-$0.6}
\expandafter\def\csname val@rev.pol.eqc.sspro.p1.uniform-fresh.linN.ci95.1\endcsname{-0.000}
\expandafter\def\csname cnt@rev.pol.eqc.sspro.p1.uniform-fresh.linN.ci95.1\endcsname{0}
\expandafter\def\csname pct@rev.pol.eqc.sspro.p1.uniform-fresh.linN.ci95.1\endcsname{0.0}
\expandafter\def\csname ppt@rev.pol.eqc.sspro.p1.uniform-fresh.linN.ci95.1\endcsname{0.0}
\expandafter\def\csname val@rev.pol.eqc.sspro.p1.uniform-fresh.linN.mean\endcsname{-0.003}
\expandafter\def\csname cnt@rev.pol.eqc.sspro.p1.uniform-fresh.linN.mean\endcsname{0}
\expandafter\def\csname pct@rev.pol.eqc.sspro.p1.uniform-fresh.linN.mean\endcsname{$-$0.3}
\expandafter\def\csname ppt@rev.pol.eqc.sspro.p1.uniform-fresh.linN.mean\endcsname{$-$0.3}
\expandafter\def\csname val@rev.pol.eqc.sspro.p1.uniform-fresh.minus.ci95.0\endcsname{-0.007}
\expandafter\def\csname cnt@rev.pol.eqc.sspro.p1.uniform-fresh.minus.ci95.0\endcsname{0}
\expandafter\def\csname pct@rev.pol.eqc.sspro.p1.uniform-fresh.minus.ci95.0\endcsname{$-$0.7}
\expandafter\def\csname ppt@rev.pol.eqc.sspro.p1.uniform-fresh.minus.ci95.0\endcsname{$-$0.7}
\expandafter\def\csname val@rev.pol.eqc.sspro.p1.uniform-fresh.minus.ci95.1\endcsname{-0.000}
\expandafter\def\csname cnt@rev.pol.eqc.sspro.p1.uniform-fresh.minus.ci95.1\endcsname{0}
\expandafter\def\csname pct@rev.pol.eqc.sspro.p1.uniform-fresh.minus.ci95.1\endcsname{0.0}
\expandafter\def\csname ppt@rev.pol.eqc.sspro.p1.uniform-fresh.minus.ci95.1\endcsname{0.0}
\expandafter\def\csname val@rev.pol.eqc.sspro.p1.uniform-fresh.minus.mean\endcsname{-0.003}
\expandafter\def\csname cnt@rev.pol.eqc.sspro.p1.uniform-fresh.minus.mean\endcsname{0}
\expandafter\def\csname pct@rev.pol.eqc.sspro.p1.uniform-fresh.minus.mean\endcsname{$-$0.3}
\expandafter\def\csname ppt@rev.pol.eqc.sspro.p1.uniform-fresh.minus.mean\endcsname{$-$0.3}
\expandafter\def\csname val@rev.pol.eqc.sspro.p1.uniform-fresh.minus.p-gt\endcsname{0.990}
\expandafter\def\csname cnt@rev.pol.eqc.sspro.p1.uniform-fresh.minus.p-gt\endcsname{1}
\expandafter\def\csname pct@rev.pol.eqc.sspro.p1.uniform-fresh.minus.p-gt\endcsname{99.0}
\expandafter\def\csname ppt@rev.pol.eqc.sspro.p1.uniform-fresh.minus.p-gt\endcsname{$+$99.0}
\expandafter\def\csname val@rev.pol.eqc.sspro.p1.uniform-fresh.p-signflip-gt\endcsname{0.957}
\expandafter\def\csname cnt@rev.pol.eqc.sspro.p1.uniform-fresh.p-signflip-gt\endcsname{1}
\expandafter\def\csname pct@rev.pol.eqc.sspro.p1.uniform-fresh.p-signflip-gt\endcsname{95.7}
\expandafter\def\csname ppt@rev.pol.eqc.sspro.p1.uniform-fresh.p-signflip-gt\endcsname{$+$95.7}
\expandafter\def\csname val@rev.pol.eqc.sspro.p1.uniform-fresh.p-signflip-lt\endcsname{0.046}
\expandafter\def\csname cnt@rev.pol.eqc.sspro.p1.uniform-fresh.p-signflip-lt\endcsname{0}
\expandafter\def\csname pct@rev.pol.eqc.sspro.p1.uniform-fresh.p-signflip-lt\endcsname{4.6}
\expandafter\def\csname ppt@rev.pol.eqc.sspro.p1.uniform-fresh.p-signflip-lt\endcsname{$+$4.6}
\expandafter\def\csname val@rev.pol.eqc.sspro.p1.uniform-fresh.score\endcsname{0.859}
\expandafter\def\csname cnt@rev.pol.eqc.sspro.p1.uniform-fresh.score\endcsname{1}
\expandafter\def\csname pct@rev.pol.eqc.sspro.p1.uniform-fresh.score\endcsname{85.9}
\expandafter\def\csname ppt@rev.pol.eqc.sspro.p1.uniform-fresh.score\endcsname{$+$85.9}
\expandafter\def\csname val@rev.pol.eqc.sspro.p2.best-single.cost\endcsname{128.839}
\expandafter\def\csname cnt@rev.pol.eqc.sspro.p2.best-single.cost\endcsname{129}
\expandafter\def\csname pct@rev.pol.eqc.sspro.p2.best-single.cost\endcsname{12883.9}
\expandafter\def\csname ppt@rev.pol.eqc.sspro.p2.best-single.cost\endcsname{$+$12883.9}
\expandafter\def\csname val@rev.pol.eqc.sspro.p2.best-single.lin.ci95.0\endcsname{-0.029}
\expandafter\def\csname cnt@rev.pol.eqc.sspro.p2.best-single.lin.ci95.0\endcsname{0}
\expandafter\def\csname pct@rev.pol.eqc.sspro.p2.best-single.lin.ci95.0\endcsname{$-$2.9}
\expandafter\def\csname ppt@rev.pol.eqc.sspro.p2.best-single.lin.ci95.0\endcsname{$-$2.9}
\expandafter\def\csname val@rev.pol.eqc.sspro.p2.best-single.lin.ci95.1\endcsname{-0.001}
\expandafter\def\csname cnt@rev.pol.eqc.sspro.p2.best-single.lin.ci95.1\endcsname{0}
\expandafter\def\csname pct@rev.pol.eqc.sspro.p2.best-single.lin.ci95.1\endcsname{$-$0.1}
\expandafter\def\csname ppt@rev.pol.eqc.sspro.p2.best-single.lin.ci95.1\endcsname{$-$0.1}
\expandafter\def\csname val@rev.pol.eqc.sspro.p2.best-single.lin.mean\endcsname{-0.015}
\expandafter\def\csname cnt@rev.pol.eqc.sspro.p2.best-single.lin.mean\endcsname{0}
\expandafter\def\csname pct@rev.pol.eqc.sspro.p2.best-single.lin.mean\endcsname{$-$1.5}
\expandafter\def\csname ppt@rev.pol.eqc.sspro.p2.best-single.lin.mean\endcsname{$-$1.5}
\expandafter\def\csname val@rev.pol.eqc.sspro.p2.best-single.linN.ci95.0\endcsname{-0.032}
\expandafter\def\csname cnt@rev.pol.eqc.sspro.p2.best-single.linN.ci95.0\endcsname{0}
\expandafter\def\csname pct@rev.pol.eqc.sspro.p2.best-single.linN.ci95.0\endcsname{$-$3.2}
\expandafter\def\csname ppt@rev.pol.eqc.sspro.p2.best-single.linN.ci95.0\endcsname{$-$3.2}
\expandafter\def\csname val@rev.pol.eqc.sspro.p2.best-single.linN.ci95.1\endcsname{-0.008}
\expandafter\def\csname cnt@rev.pol.eqc.sspro.p2.best-single.linN.ci95.1\endcsname{0}
\expandafter\def\csname pct@rev.pol.eqc.sspro.p2.best-single.linN.ci95.1\endcsname{$-$0.8}
\expandafter\def\csname ppt@rev.pol.eqc.sspro.p2.best-single.linN.ci95.1\endcsname{$-$0.8}
\expandafter\def\csname val@rev.pol.eqc.sspro.p2.best-single.linN.mean\endcsname{-0.020}
\expandafter\def\csname cnt@rev.pol.eqc.sspro.p2.best-single.linN.mean\endcsname{0}
\expandafter\def\csname pct@rev.pol.eqc.sspro.p2.best-single.linN.mean\endcsname{$-$2.0}
\expandafter\def\csname ppt@rev.pol.eqc.sspro.p2.best-single.linN.mean\endcsname{$-$2.0}
\expandafter\def\csname val@rev.pol.eqc.sspro.p2.best-single.minus.ci95.0\endcsname{-0.030}
\expandafter\def\csname cnt@rev.pol.eqc.sspro.p2.best-single.minus.ci95.0\endcsname{0}
\expandafter\def\csname pct@rev.pol.eqc.sspro.p2.best-single.minus.ci95.0\endcsname{$-$3.0}
\expandafter\def\csname ppt@rev.pol.eqc.sspro.p2.best-single.minus.ci95.0\endcsname{$-$3.0}
\expandafter\def\csname val@rev.pol.eqc.sspro.p2.best-single.minus.ci95.1\endcsname{-0.005}
\expandafter\def\csname cnt@rev.pol.eqc.sspro.p2.best-single.minus.ci95.1\endcsname{0}
\expandafter\def\csname pct@rev.pol.eqc.sspro.p2.best-single.minus.ci95.1\endcsname{$-$0.5}
\expandafter\def\csname ppt@rev.pol.eqc.sspro.p2.best-single.minus.ci95.1\endcsname{$-$0.5}
\expandafter\def\csname val@rev.pol.eqc.sspro.p2.best-single.minus.mean\endcsname{-0.018}
\expandafter\def\csname cnt@rev.pol.eqc.sspro.p2.best-single.minus.mean\endcsname{0}
\expandafter\def\csname pct@rev.pol.eqc.sspro.p2.best-single.minus.mean\endcsname{$-$1.8}
\expandafter\def\csname ppt@rev.pol.eqc.sspro.p2.best-single.minus.mean\endcsname{$-$1.8}
\expandafter\def\csname val@rev.pol.eqc.sspro.p2.best-single.minus.p-gt\endcsname{0.995}
\expandafter\def\csname cnt@rev.pol.eqc.sspro.p2.best-single.minus.p-gt\endcsname{1}
\expandafter\def\csname pct@rev.pol.eqc.sspro.p2.best-single.minus.p-gt\endcsname{99.5}
\expandafter\def\csname ppt@rev.pol.eqc.sspro.p2.best-single.minus.p-gt\endcsname{$+$99.5}
\expandafter\def\csname val@rev.pol.eqc.sspro.p2.best-single.p-signflip-gt\endcsname{0.989}
\expandafter\def\csname cnt@rev.pol.eqc.sspro.p2.best-single.p-signflip-gt\endcsname{1}
\expandafter\def\csname pct@rev.pol.eqc.sspro.p2.best-single.p-signflip-gt\endcsname{98.9}
\expandafter\def\csname ppt@rev.pol.eqc.sspro.p2.best-single.p-signflip-gt\endcsname{$+$98.9}
\expandafter\def\csname val@rev.pol.eqc.sspro.p2.best-single.p-signflip-lt\endcsname{0.012}
\expandafter\def\csname cnt@rev.pol.eqc.sspro.p2.best-single.p-signflip-lt\endcsname{0}
\expandafter\def\csname pct@rev.pol.eqc.sspro.p2.best-single.p-signflip-lt\endcsname{1.2}
\expandafter\def\csname ppt@rev.pol.eqc.sspro.p2.best-single.p-signflip-lt\endcsname{$+$1.2}
\expandafter\def\csname val@rev.pol.eqc.sspro.p2.best-single.score\endcsname{0.909}
\expandafter\def\csname cnt@rev.pol.eqc.sspro.p2.best-single.score\endcsname{1}
\expandafter\def\csname pct@rev.pol.eqc.sspro.p2.best-single.score\endcsname{90.9}
\expandafter\def\csname ppt@rev.pol.eqc.sspro.p2.best-single.score\endcsname{$+$90.9}
\expandafter\def\csname val@rev.pol.eqc.sspro.p2.generic-ppi-source.cost\endcsname{128.839}
\expandafter\def\csname cnt@rev.pol.eqc.sspro.p2.generic-ppi-source.cost\endcsname{129}
\expandafter\def\csname pct@rev.pol.eqc.sspro.p2.generic-ppi-source.cost\endcsname{12883.9}
\expandafter\def\csname ppt@rev.pol.eqc.sspro.p2.generic-ppi-source.cost\endcsname{$+$12883.9}
\expandafter\def\csname val@rev.pol.eqc.sspro.p2.generic-ppi-source.lin.ci95.0\endcsname{-0.008}
\expandafter\def\csname cnt@rev.pol.eqc.sspro.p2.generic-ppi-source.lin.ci95.0\endcsname{0}
\expandafter\def\csname pct@rev.pol.eqc.sspro.p2.generic-ppi-source.lin.ci95.0\endcsname{$-$0.8}
\expandafter\def\csname ppt@rev.pol.eqc.sspro.p2.generic-ppi-source.lin.ci95.0\endcsname{$-$0.8}
\expandafter\def\csname val@rev.pol.eqc.sspro.p2.generic-ppi-source.lin.ci95.1\endcsname{-0.003}
\expandafter\def\csname cnt@rev.pol.eqc.sspro.p2.generic-ppi-source.lin.ci95.1\endcsname{0}
\expandafter\def\csname pct@rev.pol.eqc.sspro.p2.generic-ppi-source.lin.ci95.1\endcsname{$-$0.3}
\expandafter\def\csname ppt@rev.pol.eqc.sspro.p2.generic-ppi-source.lin.ci95.1\endcsname{$-$0.3}
\expandafter\def\csname val@rev.pol.eqc.sspro.p2.generic-ppi-source.lin.mean\endcsname{-0.005}
\expandafter\def\csname cnt@rev.pol.eqc.sspro.p2.generic-ppi-source.lin.mean\endcsname{0}
\expandafter\def\csname pct@rev.pol.eqc.sspro.p2.generic-ppi-source.lin.mean\endcsname{$-$0.5}
\expandafter\def\csname ppt@rev.pol.eqc.sspro.p2.generic-ppi-source.lin.mean\endcsname{$-$0.5}
\expandafter\def\csname val@rev.pol.eqc.sspro.p2.generic-ppi-source.linN.ci95.0\endcsname{-0.018}
\expandafter\def\csname cnt@rev.pol.eqc.sspro.p2.generic-ppi-source.linN.ci95.0\endcsname{0}
\expandafter\def\csname pct@rev.pol.eqc.sspro.p2.generic-ppi-source.linN.ci95.0\endcsname{$-$1.8}
\expandafter\def\csname ppt@rev.pol.eqc.sspro.p2.generic-ppi-source.linN.ci95.0\endcsname{$-$1.8}
\expandafter\def\csname val@rev.pol.eqc.sspro.p2.generic-ppi-source.linN.ci95.1\endcsname{-0.006}
\expandafter\def\csname cnt@rev.pol.eqc.sspro.p2.generic-ppi-source.linN.ci95.1\endcsname{0}
\expandafter\def\csname pct@rev.pol.eqc.sspro.p2.generic-ppi-source.linN.ci95.1\endcsname{$-$0.6}
\expandafter\def\csname ppt@rev.pol.eqc.sspro.p2.generic-ppi-source.linN.ci95.1\endcsname{$-$0.6}
\expandafter\def\csname val@rev.pol.eqc.sspro.p2.generic-ppi-source.linN.mean\endcsname{-0.012}
\expandafter\def\csname cnt@rev.pol.eqc.sspro.p2.generic-ppi-source.linN.mean\endcsname{0}
\expandafter\def\csname pct@rev.pol.eqc.sspro.p2.generic-ppi-source.linN.mean\endcsname{$-$1.2}
\expandafter\def\csname ppt@rev.pol.eqc.sspro.p2.generic-ppi-source.linN.mean\endcsname{$-$1.2}
\expandafter\def\csname val@rev.pol.eqc.sspro.p2.generic-ppi-source.minus.ci95.0\endcsname{-0.017}
\expandafter\def\csname cnt@rev.pol.eqc.sspro.p2.generic-ppi-source.minus.ci95.0\endcsname{0}
\expandafter\def\csname pct@rev.pol.eqc.sspro.p2.generic-ppi-source.minus.ci95.0\endcsname{$-$1.7}
\expandafter\def\csname ppt@rev.pol.eqc.sspro.p2.generic-ppi-source.minus.ci95.0\endcsname{$-$1.7}
\expandafter\def\csname val@rev.pol.eqc.sspro.p2.generic-ppi-source.minus.ci95.1\endcsname{-0.006}
\expandafter\def\csname cnt@rev.pol.eqc.sspro.p2.generic-ppi-source.minus.ci95.1\endcsname{0}
\expandafter\def\csname pct@rev.pol.eqc.sspro.p2.generic-ppi-source.minus.ci95.1\endcsname{$-$0.6}
\expandafter\def\csname ppt@rev.pol.eqc.sspro.p2.generic-ppi-source.minus.ci95.1\endcsname{$-$0.6}
\expandafter\def\csname val@rev.pol.eqc.sspro.p2.generic-ppi-source.minus.mean\endcsname{-0.011}
\expandafter\def\csname cnt@rev.pol.eqc.sspro.p2.generic-ppi-source.minus.mean\endcsname{0}
\expandafter\def\csname pct@rev.pol.eqc.sspro.p2.generic-ppi-source.minus.mean\endcsname{$-$1.1}
\expandafter\def\csname ppt@rev.pol.eqc.sspro.p2.generic-ppi-source.minus.mean\endcsname{$-$1.1}
\expandafter\def\csname val@rev.pol.eqc.sspro.p2.generic-ppi-source.minus.p-gt\endcsname{1.000}
\expandafter\def\csname cnt@rev.pol.eqc.sspro.p2.generic-ppi-source.minus.p-gt\endcsname{1}
\expandafter\def\csname pct@rev.pol.eqc.sspro.p2.generic-ppi-source.minus.p-gt\endcsname{100.0}
\expandafter\def\csname ppt@rev.pol.eqc.sspro.p2.generic-ppi-source.minus.p-gt\endcsname{$+$100.0}
\expandafter\def\csname val@rev.pol.eqc.sspro.p2.generic-ppi-source.p-signflip-gt\endcsname{1.000}
\expandafter\def\csname cnt@rev.pol.eqc.sspro.p2.generic-ppi-source.p-signflip-gt\endcsname{1}
\expandafter\def\csname pct@rev.pol.eqc.sspro.p2.generic-ppi-source.p-signflip-gt\endcsname{100.0}
\expandafter\def\csname ppt@rev.pol.eqc.sspro.p2.generic-ppi-source.p-signflip-gt\endcsname{$+$100.0}
\expandafter\def\csname val@rev.pol.eqc.sspro.p2.generic-ppi-source.p-signflip-lt\endcsname{0.001}
\expandafter\def\csname cnt@rev.pol.eqc.sspro.p2.generic-ppi-source.p-signflip-lt\endcsname{0}
\expandafter\def\csname pct@rev.pol.eqc.sspro.p2.generic-ppi-source.p-signflip-lt\endcsname{0.1}
\expandafter\def\csname ppt@rev.pol.eqc.sspro.p2.generic-ppi-source.p-signflip-lt\endcsname{$+$0.1}
\expandafter\def\csname val@rev.pol.eqc.sspro.p2.generic-ppi-source.score\endcsname{0.902}
\expandafter\def\csname cnt@rev.pol.eqc.sspro.p2.generic-ppi-source.score\endcsname{1}
\expandafter\def\csname pct@rev.pol.eqc.sspro.p2.generic-ppi-source.score\endcsname{90.2}
\expandafter\def\csname ppt@rev.pol.eqc.sspro.p2.generic-ppi-source.score\endcsname{$+$90.2}
\expandafter\def\csname val@rev.pol.eqc.sspro.p2.rcr-cost\endcsname{128.845}
\expandafter\def\csname cnt@rev.pol.eqc.sspro.p2.rcr-cost\endcsname{129}
\expandafter\def\csname pct@rev.pol.eqc.sspro.p2.rcr-cost\endcsname{12884.5}
\expandafter\def\csname ppt@rev.pol.eqc.sspro.p2.rcr-cost\endcsname{$+$12884.5}
\expandafter\def\csname val@rev.pol.eqc.sspro.p2.replay-ppi-uniform.cost\endcsname{128.845}
\expandafter\def\csname cnt@rev.pol.eqc.sspro.p2.replay-ppi-uniform.cost\endcsname{129}
\expandafter\def\csname pct@rev.pol.eqc.sspro.p2.replay-ppi-uniform.cost\endcsname{12884.5}
\expandafter\def\csname ppt@rev.pol.eqc.sspro.p2.replay-ppi-uniform.cost\endcsname{$+$12884.5}
\expandafter\def\csname val@rev.pol.eqc.sspro.p2.replay-ppi-uniform.lin.ci95.0\endcsname{0.005}
\expandafter\def\csname cnt@rev.pol.eqc.sspro.p2.replay-ppi-uniform.lin.ci95.0\endcsname{0}
\expandafter\def\csname pct@rev.pol.eqc.sspro.p2.replay-ppi-uniform.lin.ci95.0\endcsname{0.5}
\expandafter\def\csname ppt@rev.pol.eqc.sspro.p2.replay-ppi-uniform.lin.ci95.0\endcsname{$+$0.5}
\expandafter\def\csname val@rev.pol.eqc.sspro.p2.replay-ppi-uniform.lin.ci95.1\endcsname{0.014}
\expandafter\def\csname cnt@rev.pol.eqc.sspro.p2.replay-ppi-uniform.lin.ci95.1\endcsname{0}
\expandafter\def\csname pct@rev.pol.eqc.sspro.p2.replay-ppi-uniform.lin.ci95.1\endcsname{1.4}
\expandafter\def\csname ppt@rev.pol.eqc.sspro.p2.replay-ppi-uniform.lin.ci95.1\endcsname{$+$1.4}
\expandafter\def\csname val@rev.pol.eqc.sspro.p2.replay-ppi-uniform.lin.mean\endcsname{0.010}
\expandafter\def\csname cnt@rev.pol.eqc.sspro.p2.replay-ppi-uniform.lin.mean\endcsname{0}
\expandafter\def\csname pct@rev.pol.eqc.sspro.p2.replay-ppi-uniform.lin.mean\endcsname{1.0}
\expandafter\def\csname ppt@rev.pol.eqc.sspro.p2.replay-ppi-uniform.lin.mean\endcsname{$+$1.0}
\expandafter\def\csname val@rev.pol.eqc.sspro.p2.replay-ppi-uniform.linN.ci95.0\endcsname{-0.001}
\expandafter\def\csname cnt@rev.pol.eqc.sspro.p2.replay-ppi-uniform.linN.ci95.0\endcsname{0}
\expandafter\def\csname pct@rev.pol.eqc.sspro.p2.replay-ppi-uniform.linN.ci95.0\endcsname{$-$0.1}
\expandafter\def\csname ppt@rev.pol.eqc.sspro.p2.replay-ppi-uniform.linN.ci95.0\endcsname{$-$0.1}
\expandafter\def\csname val@rev.pol.eqc.sspro.p2.replay-ppi-uniform.linN.ci95.1\endcsname{0.001}
\expandafter\def\csname cnt@rev.pol.eqc.sspro.p2.replay-ppi-uniform.linN.ci95.1\endcsname{0}
\expandafter\def\csname pct@rev.pol.eqc.sspro.p2.replay-ppi-uniform.linN.ci95.1\endcsname{0.1}
\expandafter\def\csname ppt@rev.pol.eqc.sspro.p2.replay-ppi-uniform.linN.ci95.1\endcsname{$+$0.1}
\expandafter\def\csname val@rev.pol.eqc.sspro.p2.replay-ppi-uniform.linN.mean\endcsname{0.000}
\expandafter\def\csname cnt@rev.pol.eqc.sspro.p2.replay-ppi-uniform.linN.mean\endcsname{0}
\expandafter\def\csname pct@rev.pol.eqc.sspro.p2.replay-ppi-uniform.linN.mean\endcsname{0.0}
\expandafter\def\csname ppt@rev.pol.eqc.sspro.p2.replay-ppi-uniform.linN.mean\endcsname{0.0}
\expandafter\def\csname val@rev.pol.eqc.sspro.p2.replay-ppi-uniform.minus.ci95.0\endcsname{-0.001}
\expandafter\def\csname cnt@rev.pol.eqc.sspro.p2.replay-ppi-uniform.minus.ci95.0\endcsname{0}
\expandafter\def\csname pct@rev.pol.eqc.sspro.p2.replay-ppi-uniform.minus.ci95.0\endcsname{$-$0.1}
\expandafter\def\csname ppt@rev.pol.eqc.sspro.p2.replay-ppi-uniform.minus.ci95.0\endcsname{$-$0.1}
\expandafter\def\csname val@rev.pol.eqc.sspro.p2.replay-ppi-uniform.minus.ci95.1\endcsname{0.001}
\expandafter\def\csname cnt@rev.pol.eqc.sspro.p2.replay-ppi-uniform.minus.ci95.1\endcsname{0}
\expandafter\def\csname pct@rev.pol.eqc.sspro.p2.replay-ppi-uniform.minus.ci95.1\endcsname{0.1}
\expandafter\def\csname ppt@rev.pol.eqc.sspro.p2.replay-ppi-uniform.minus.ci95.1\endcsname{$+$0.1}
\expandafter\def\csname val@rev.pol.eqc.sspro.p2.replay-ppi-uniform.minus.mean\endcsname{0.000}
\expandafter\def\csname cnt@rev.pol.eqc.sspro.p2.replay-ppi-uniform.minus.mean\endcsname{0}
\expandafter\def\csname pct@rev.pol.eqc.sspro.p2.replay-ppi-uniform.minus.mean\endcsname{0.0}
\expandafter\def\csname ppt@rev.pol.eqc.sspro.p2.replay-ppi-uniform.minus.mean\endcsname{0.0}
\expandafter\def\csname val@rev.pol.eqc.sspro.p2.replay-ppi-uniform.minus.p-gt\endcsname{0.434}
\expandafter\def\csname cnt@rev.pol.eqc.sspro.p2.replay-ppi-uniform.minus.p-gt\endcsname{0}
\expandafter\def\csname pct@rev.pol.eqc.sspro.p2.replay-ppi-uniform.minus.p-gt\endcsname{43.4}
\expandafter\def\csname ppt@rev.pol.eqc.sspro.p2.replay-ppi-uniform.minus.p-gt\endcsname{$+$43.4}
\expandafter\def\csname val@rev.pol.eqc.sspro.p2.replay-ppi-uniform.p-signflip-gt\endcsname{0.500}
\expandafter\def\csname cnt@rev.pol.eqc.sspro.p2.replay-ppi-uniform.p-signflip-gt\endcsname{0}
\expandafter\def\csname pct@rev.pol.eqc.sspro.p2.replay-ppi-uniform.p-signflip-gt\endcsname{50.0}
\expandafter\def\csname ppt@rev.pol.eqc.sspro.p2.replay-ppi-uniform.p-signflip-gt\endcsname{$+$50.0}
\expandafter\def\csname val@rev.pol.eqc.sspro.p2.replay-ppi-uniform.p-signflip-lt\endcsname{0.501}
\expandafter\def\csname cnt@rev.pol.eqc.sspro.p2.replay-ppi-uniform.p-signflip-lt\endcsname{1}
\expandafter\def\csname pct@rev.pol.eqc.sspro.p2.replay-ppi-uniform.p-signflip-lt\endcsname{50.1}
\expandafter\def\csname ppt@rev.pol.eqc.sspro.p2.replay-ppi-uniform.p-signflip-lt\endcsname{$+$50.1}
\expandafter\def\csname val@rev.pol.eqc.sspro.p2.replay-ppi-uniform.score\endcsname{0.891}
\expandafter\def\csname cnt@rev.pol.eqc.sspro.p2.replay-ppi-uniform.score\endcsname{1}
\expandafter\def\csname pct@rev.pol.eqc.sspro.p2.replay-ppi-uniform.score\endcsname{89.1}
\expandafter\def\csname ppt@rev.pol.eqc.sspro.p2.replay-ppi-uniform.score\endcsname{$+$89.1}
\expandafter\def\csname val@rev.pol.eqc.sspro.p2.uniform-fresh.cost\endcsname{128.839}
\expandafter\def\csname cnt@rev.pol.eqc.sspro.p2.uniform-fresh.cost\endcsname{129}
\expandafter\def\csname pct@rev.pol.eqc.sspro.p2.uniform-fresh.cost\endcsname{12883.9}
\expandafter\def\csname ppt@rev.pol.eqc.sspro.p2.uniform-fresh.cost\endcsname{$+$12883.9}
\expandafter\def\csname val@rev.pol.eqc.sspro.p2.uniform-fresh.lin.ci95.0\endcsname{-0.013}
\expandafter\def\csname cnt@rev.pol.eqc.sspro.p2.uniform-fresh.lin.ci95.0\endcsname{0}
\expandafter\def\csname pct@rev.pol.eqc.sspro.p2.uniform-fresh.lin.ci95.0\endcsname{$-$1.3}
\expandafter\def\csname ppt@rev.pol.eqc.sspro.p2.uniform-fresh.lin.ci95.0\endcsname{$-$1.3}
\expandafter\def\csname val@rev.pol.eqc.sspro.p2.uniform-fresh.lin.ci95.1\endcsname{-0.005}
\expandafter\def\csname cnt@rev.pol.eqc.sspro.p2.uniform-fresh.lin.ci95.1\endcsname{0}
\expandafter\def\csname pct@rev.pol.eqc.sspro.p2.uniform-fresh.lin.ci95.1\endcsname{$-$0.5}
\expandafter\def\csname ppt@rev.pol.eqc.sspro.p2.uniform-fresh.lin.ci95.1\endcsname{$-$0.5}
\expandafter\def\csname val@rev.pol.eqc.sspro.p2.uniform-fresh.lin.mean\endcsname{-0.008}
\expandafter\def\csname cnt@rev.pol.eqc.sspro.p2.uniform-fresh.lin.mean\endcsname{0}
\expandafter\def\csname pct@rev.pol.eqc.sspro.p2.uniform-fresh.lin.mean\endcsname{$-$0.8}
\expandafter\def\csname ppt@rev.pol.eqc.sspro.p2.uniform-fresh.lin.mean\endcsname{$-$0.8}
\expandafter\def\csname val@rev.pol.eqc.sspro.p2.uniform-fresh.linN.ci95.0\endcsname{-0.017}
\expandafter\def\csname cnt@rev.pol.eqc.sspro.p2.uniform-fresh.linN.ci95.0\endcsname{0}
\expandafter\def\csname pct@rev.pol.eqc.sspro.p2.uniform-fresh.linN.ci95.0\endcsname{$-$1.7}
\expandafter\def\csname ppt@rev.pol.eqc.sspro.p2.uniform-fresh.linN.ci95.0\endcsname{$-$1.7}
\expandafter\def\csname val@rev.pol.eqc.sspro.p2.uniform-fresh.linN.ci95.1\endcsname{-0.006}
\expandafter\def\csname cnt@rev.pol.eqc.sspro.p2.uniform-fresh.linN.ci95.1\endcsname{0}
\expandafter\def\csname pct@rev.pol.eqc.sspro.p2.uniform-fresh.linN.ci95.1\endcsname{$-$0.6}
\expandafter\def\csname ppt@rev.pol.eqc.sspro.p2.uniform-fresh.linN.ci95.1\endcsname{$-$0.6}
\expandafter\def\csname val@rev.pol.eqc.sspro.p2.uniform-fresh.linN.mean\endcsname{-0.011}
\expandafter\def\csname cnt@rev.pol.eqc.sspro.p2.uniform-fresh.linN.mean\endcsname{0}
\expandafter\def\csname pct@rev.pol.eqc.sspro.p2.uniform-fresh.linN.mean\endcsname{$-$1.1}
\expandafter\def\csname ppt@rev.pol.eqc.sspro.p2.uniform-fresh.linN.mean\endcsname{$-$1.1}
\expandafter\def\csname val@rev.pol.eqc.sspro.p2.uniform-fresh.minus.ci95.0\endcsname{-0.017}
\expandafter\def\csname cnt@rev.pol.eqc.sspro.p2.uniform-fresh.minus.ci95.0\endcsname{0}
\expandafter\def\csname pct@rev.pol.eqc.sspro.p2.uniform-fresh.minus.ci95.0\endcsname{$-$1.7}
\expandafter\def\csname ppt@rev.pol.eqc.sspro.p2.uniform-fresh.minus.ci95.0\endcsname{$-$1.7}
\expandafter\def\csname val@rev.pol.eqc.sspro.p2.uniform-fresh.minus.ci95.1\endcsname{-0.006}
\expandafter\def\csname cnt@rev.pol.eqc.sspro.p2.uniform-fresh.minus.ci95.1\endcsname{0}
\expandafter\def\csname pct@rev.pol.eqc.sspro.p2.uniform-fresh.minus.ci95.1\endcsname{$-$0.6}
\expandafter\def\csname ppt@rev.pol.eqc.sspro.p2.uniform-fresh.minus.ci95.1\endcsname{$-$0.6}
\expandafter\def\csname val@rev.pol.eqc.sspro.p2.uniform-fresh.minus.mean\endcsname{-0.011}
\expandafter\def\csname cnt@rev.pol.eqc.sspro.p2.uniform-fresh.minus.mean\endcsname{0}
\expandafter\def\csname pct@rev.pol.eqc.sspro.p2.uniform-fresh.minus.mean\endcsname{$-$1.1}
\expandafter\def\csname ppt@rev.pol.eqc.sspro.p2.uniform-fresh.minus.mean\endcsname{$-$1.1}
\expandafter\def\csname val@rev.pol.eqc.sspro.p2.uniform-fresh.minus.p-gt\endcsname{1.000}
\expandafter\def\csname cnt@rev.pol.eqc.sspro.p2.uniform-fresh.minus.p-gt\endcsname{1}
\expandafter\def\csname pct@rev.pol.eqc.sspro.p2.uniform-fresh.minus.p-gt\endcsname{100.0}
\expandafter\def\csname ppt@rev.pol.eqc.sspro.p2.uniform-fresh.minus.p-gt\endcsname{$+$100.0}
\expandafter\def\csname val@rev.pol.eqc.sspro.p2.uniform-fresh.p-signflip-gt\endcsname{1.000}
\expandafter\def\csname cnt@rev.pol.eqc.sspro.p2.uniform-fresh.p-signflip-gt\endcsname{1}
\expandafter\def\csname pct@rev.pol.eqc.sspro.p2.uniform-fresh.p-signflip-gt\endcsname{100.0}
\expandafter\def\csname ppt@rev.pol.eqc.sspro.p2.uniform-fresh.p-signflip-gt\endcsname{$+$100.0}
\expandafter\def\csname val@rev.pol.eqc.sspro.p2.uniform-fresh.p-signflip-lt\endcsname{0.001}
\expandafter\def\csname cnt@rev.pol.eqc.sspro.p2.uniform-fresh.p-signflip-lt\endcsname{0}
\expandafter\def\csname pct@rev.pol.eqc.sspro.p2.uniform-fresh.p-signflip-lt\endcsname{0.1}
\expandafter\def\csname ppt@rev.pol.eqc.sspro.p2.uniform-fresh.p-signflip-lt\endcsname{$+$0.1}
\expandafter\def\csname val@rev.pol.eqc.sspro.p2.uniform-fresh.score\endcsname{0.902}
\expandafter\def\csname cnt@rev.pol.eqc.sspro.p2.uniform-fresh.score\endcsname{1}
\expandafter\def\csname pct@rev.pol.eqc.sspro.p2.uniform-fresh.score\endcsname{90.2}
\expandafter\def\csname ppt@rev.pol.eqc.sspro.p2.uniform-fresh.score\endcsname{$+$90.2}
\expandafter\def\csname val@rev.pol.eqc.uiv.p1.best-single.cost\endcsname{257.120}
\expandafter\def\csname cnt@rev.pol.eqc.uiv.p1.best-single.cost\endcsname{257}
\expandafter\def\csname pct@rev.pol.eqc.uiv.p1.best-single.cost\endcsname{25712.0}
\expandafter\def\csname ppt@rev.pol.eqc.uiv.p1.best-single.cost\endcsname{$+$25712.0}
\expandafter\def\csname val@rev.pol.eqc.uiv.p1.best-single.lin.ci95.0\endcsname{-0.022}
\expandafter\def\csname cnt@rev.pol.eqc.uiv.p1.best-single.lin.ci95.0\endcsname{0}
\expandafter\def\csname pct@rev.pol.eqc.uiv.p1.best-single.lin.ci95.0\endcsname{$-$2.2}
\expandafter\def\csname ppt@rev.pol.eqc.uiv.p1.best-single.lin.ci95.0\endcsname{$-$2.2}
\expandafter\def\csname val@rev.pol.eqc.uiv.p1.best-single.lin.ci95.1\endcsname{-0.008}
\expandafter\def\csname cnt@rev.pol.eqc.uiv.p1.best-single.lin.ci95.1\endcsname{0}
\expandafter\def\csname pct@rev.pol.eqc.uiv.p1.best-single.lin.ci95.1\endcsname{$-$0.8}
\expandafter\def\csname ppt@rev.pol.eqc.uiv.p1.best-single.lin.ci95.1\endcsname{$-$0.8}
\expandafter\def\csname val@rev.pol.eqc.uiv.p1.best-single.lin.mean\endcsname{-0.015}
\expandafter\def\csname cnt@rev.pol.eqc.uiv.p1.best-single.lin.mean\endcsname{0}
\expandafter\def\csname pct@rev.pol.eqc.uiv.p1.best-single.lin.mean\endcsname{$-$1.5}
\expandafter\def\csname ppt@rev.pol.eqc.uiv.p1.best-single.lin.mean\endcsname{$-$1.5}
\expandafter\def\csname val@rev.pol.eqc.uiv.p1.best-single.linN.ci95.0\endcsname{-0.009}
\expandafter\def\csname cnt@rev.pol.eqc.uiv.p1.best-single.linN.ci95.0\endcsname{0}
\expandafter\def\csname pct@rev.pol.eqc.uiv.p1.best-single.linN.ci95.0\endcsname{$-$0.9}
\expandafter\def\csname ppt@rev.pol.eqc.uiv.p1.best-single.linN.ci95.0\endcsname{$-$0.9}
\expandafter\def\csname val@rev.pol.eqc.uiv.p1.best-single.linN.ci95.1\endcsname{0.002}
\expandafter\def\csname cnt@rev.pol.eqc.uiv.p1.best-single.linN.ci95.1\endcsname{0}
\expandafter\def\csname pct@rev.pol.eqc.uiv.p1.best-single.linN.ci95.1\endcsname{0.2}
\expandafter\def\csname ppt@rev.pol.eqc.uiv.p1.best-single.linN.ci95.1\endcsname{$+$0.2}
\expandafter\def\csname val@rev.pol.eqc.uiv.p1.best-single.linN.mean\endcsname{-0.004}
\expandafter\def\csname cnt@rev.pol.eqc.uiv.p1.best-single.linN.mean\endcsname{0}
\expandafter\def\csname pct@rev.pol.eqc.uiv.p1.best-single.linN.mean\endcsname{$-$0.4}
\expandafter\def\csname ppt@rev.pol.eqc.uiv.p1.best-single.linN.mean\endcsname{$-$0.4}
\expandafter\def\csname val@rev.pol.eqc.uiv.p1.best-single.minus.ci95.0\endcsname{0.004}
\expandafter\def\csname cnt@rev.pol.eqc.uiv.p1.best-single.minus.ci95.0\endcsname{0}
\expandafter\def\csname pct@rev.pol.eqc.uiv.p1.best-single.minus.ci95.0\endcsname{0.4}
\expandafter\def\csname ppt@rev.pol.eqc.uiv.p1.best-single.minus.ci95.0\endcsname{$+$0.4}
\expandafter\def\csname val@rev.pol.eqc.uiv.p1.best-single.minus.ci95.1\endcsname{0.028}
\expandafter\def\csname cnt@rev.pol.eqc.uiv.p1.best-single.minus.ci95.1\endcsname{0}
\expandafter\def\csname pct@rev.pol.eqc.uiv.p1.best-single.minus.ci95.1\endcsname{2.8}
\expandafter\def\csname ppt@rev.pol.eqc.uiv.p1.best-single.minus.ci95.1\endcsname{$+$2.8}
\expandafter\def\csname val@rev.pol.eqc.uiv.p1.best-single.minus.mean\endcsname{0.016}
\expandafter\def\csname cnt@rev.pol.eqc.uiv.p1.best-single.minus.mean\endcsname{0}
\expandafter\def\csname pct@rev.pol.eqc.uiv.p1.best-single.minus.mean\endcsname{1.6}
\expandafter\def\csname ppt@rev.pol.eqc.uiv.p1.best-single.minus.mean\endcsname{$+$1.6}
\expandafter\def\csname val@rev.pol.eqc.uiv.p1.best-single.minus.p-gt\endcsname{0.003}
\expandafter\def\csname cnt@rev.pol.eqc.uiv.p1.best-single.minus.p-gt\endcsname{0}
\expandafter\def\csname pct@rev.pol.eqc.uiv.p1.best-single.minus.p-gt\endcsname{0.3}
\expandafter\def\csname ppt@rev.pol.eqc.uiv.p1.best-single.minus.p-gt\endcsname{$+$0.3}
\expandafter\def\csname val@rev.pol.eqc.uiv.p1.best-single.p-signflip-gt\endcsname{0.007}
\expandafter\def\csname cnt@rev.pol.eqc.uiv.p1.best-single.p-signflip-gt\endcsname{0}
\expandafter\def\csname pct@rev.pol.eqc.uiv.p1.best-single.p-signflip-gt\endcsname{0.7}
\expandafter\def\csname ppt@rev.pol.eqc.uiv.p1.best-single.p-signflip-gt\endcsname{$+$0.7}
\expandafter\def\csname val@rev.pol.eqc.uiv.p1.best-single.p-signflip-lt\endcsname{0.993}
\expandafter\def\csname cnt@rev.pol.eqc.uiv.p1.best-single.p-signflip-lt\endcsname{1}
\expandafter\def\csname pct@rev.pol.eqc.uiv.p1.best-single.p-signflip-lt\endcsname{99.3}
\expandafter\def\csname ppt@rev.pol.eqc.uiv.p1.best-single.p-signflip-lt\endcsname{$+$99.3}
\expandafter\def\csname val@rev.pol.eqc.uiv.p1.best-single.score\endcsname{0.788}
\expandafter\def\csname cnt@rev.pol.eqc.uiv.p1.best-single.score\endcsname{1}
\expandafter\def\csname pct@rev.pol.eqc.uiv.p1.best-single.score\endcsname{78.8}
\expandafter\def\csname ppt@rev.pol.eqc.uiv.p1.best-single.score\endcsname{$+$78.8}
\expandafter\def\csname val@rev.pol.eqc.uiv.p1.generic-ppi-source.cost\endcsname{257.120}
\expandafter\def\csname cnt@rev.pol.eqc.uiv.p1.generic-ppi-source.cost\endcsname{257}
\expandafter\def\csname pct@rev.pol.eqc.uiv.p1.generic-ppi-source.cost\endcsname{25712.0}
\expandafter\def\csname ppt@rev.pol.eqc.uiv.p1.generic-ppi-source.cost\endcsname{$+$25712.0}
\expandafter\def\csname val@rev.pol.eqc.uiv.p1.generic-ppi-source.lin.ci95.0\endcsname{0.003}
\expandafter\def\csname cnt@rev.pol.eqc.uiv.p1.generic-ppi-source.lin.ci95.0\endcsname{0}
\expandafter\def\csname pct@rev.pol.eqc.uiv.p1.generic-ppi-source.lin.ci95.0\endcsname{0.3}
\expandafter\def\csname ppt@rev.pol.eqc.uiv.p1.generic-ppi-source.lin.ci95.0\endcsname{$+$0.3}
\expandafter\def\csname val@rev.pol.eqc.uiv.p1.generic-ppi-source.lin.ci95.1\endcsname{0.007}
\expandafter\def\csname cnt@rev.pol.eqc.uiv.p1.generic-ppi-source.lin.ci95.1\endcsname{0}
\expandafter\def\csname pct@rev.pol.eqc.uiv.p1.generic-ppi-source.lin.ci95.1\endcsname{0.7}
\expandafter\def\csname ppt@rev.pol.eqc.uiv.p1.generic-ppi-source.lin.ci95.1\endcsname{$+$0.7}
\expandafter\def\csname val@rev.pol.eqc.uiv.p1.generic-ppi-source.lin.mean\endcsname{0.005}
\expandafter\def\csname cnt@rev.pol.eqc.uiv.p1.generic-ppi-source.lin.mean\endcsname{0}
\expandafter\def\csname pct@rev.pol.eqc.uiv.p1.generic-ppi-source.lin.mean\endcsname{0.5}
\expandafter\def\csname ppt@rev.pol.eqc.uiv.p1.generic-ppi-source.lin.mean\endcsname{$+$0.5}
\expandafter\def\csname val@rev.pol.eqc.uiv.p1.generic-ppi-source.linN.ci95.0\endcsname{0.004}
\expandafter\def\csname cnt@rev.pol.eqc.uiv.p1.generic-ppi-source.linN.ci95.0\endcsname{0}
\expandafter\def\csname pct@rev.pol.eqc.uiv.p1.generic-ppi-source.linN.ci95.0\endcsname{0.4}
\expandafter\def\csname ppt@rev.pol.eqc.uiv.p1.generic-ppi-source.linN.ci95.0\endcsname{$+$0.4}
\expandafter\def\csname val@rev.pol.eqc.uiv.p1.generic-ppi-source.linN.ci95.1\endcsname{0.010}
\expandafter\def\csname cnt@rev.pol.eqc.uiv.p1.generic-ppi-source.linN.ci95.1\endcsname{0}
\expandafter\def\csname pct@rev.pol.eqc.uiv.p1.generic-ppi-source.linN.ci95.1\endcsname{1.0}
\expandafter\def\csname ppt@rev.pol.eqc.uiv.p1.generic-ppi-source.linN.ci95.1\endcsname{$+$1.0}
\expandafter\def\csname val@rev.pol.eqc.uiv.p1.generic-ppi-source.linN.mean\endcsname{0.007}
\expandafter\def\csname cnt@rev.pol.eqc.uiv.p1.generic-ppi-source.linN.mean\endcsname{0}
\expandafter\def\csname pct@rev.pol.eqc.uiv.p1.generic-ppi-source.linN.mean\endcsname{0.7}
\expandafter\def\csname ppt@rev.pol.eqc.uiv.p1.generic-ppi-source.linN.mean\endcsname{$+$0.7}
\expandafter\def\csname val@rev.pol.eqc.uiv.p1.generic-ppi-source.minus.ci95.0\endcsname{-0.001}
\expandafter\def\csname cnt@rev.pol.eqc.uiv.p1.generic-ppi-source.minus.ci95.0\endcsname{0}
\expandafter\def\csname pct@rev.pol.eqc.uiv.p1.generic-ppi-source.minus.ci95.0\endcsname{$-$0.1}
\expandafter\def\csname ppt@rev.pol.eqc.uiv.p1.generic-ppi-source.minus.ci95.0\endcsname{$-$0.1}
\expandafter\def\csname val@rev.pol.eqc.uiv.p1.generic-ppi-source.minus.ci95.1\endcsname{0.001}
\expandafter\def\csname cnt@rev.pol.eqc.uiv.p1.generic-ppi-source.minus.ci95.1\endcsname{0}
\expandafter\def\csname pct@rev.pol.eqc.uiv.p1.generic-ppi-source.minus.ci95.1\endcsname{0.1}
\expandafter\def\csname ppt@rev.pol.eqc.uiv.p1.generic-ppi-source.minus.ci95.1\endcsname{$+$0.1}
\expandafter\def\csname val@rev.pol.eqc.uiv.p1.generic-ppi-source.minus.mean\endcsname{0.000}
\expandafter\def\csname cnt@rev.pol.eqc.uiv.p1.generic-ppi-source.minus.mean\endcsname{0}
\expandafter\def\csname pct@rev.pol.eqc.uiv.p1.generic-ppi-source.minus.mean\endcsname{0.0}
\expandafter\def\csname ppt@rev.pol.eqc.uiv.p1.generic-ppi-source.minus.mean\endcsname{0.0}
\expandafter\def\csname val@rev.pol.eqc.uiv.p1.generic-ppi-source.minus.p-gt\endcsname{0.377}
\expandafter\def\csname cnt@rev.pol.eqc.uiv.p1.generic-ppi-source.minus.p-gt\endcsname{0}
\expandafter\def\csname pct@rev.pol.eqc.uiv.p1.generic-ppi-source.minus.p-gt\endcsname{37.7}
\expandafter\def\csname ppt@rev.pol.eqc.uiv.p1.generic-ppi-source.minus.p-gt\endcsname{$+$37.7}
\expandafter\def\csname val@rev.pol.eqc.uiv.p1.generic-ppi-source.p-signflip-gt\endcsname{0.417}
\expandafter\def\csname cnt@rev.pol.eqc.uiv.p1.generic-ppi-source.p-signflip-gt\endcsname{0}
\expandafter\def\csname pct@rev.pol.eqc.uiv.p1.generic-ppi-source.p-signflip-gt\endcsname{41.7}
\expandafter\def\csname ppt@rev.pol.eqc.uiv.p1.generic-ppi-source.p-signflip-gt\endcsname{$+$41.7}
\expandafter\def\csname val@rev.pol.eqc.uiv.p1.generic-ppi-source.p-signflip-lt\endcsname{0.583}
\expandafter\def\csname cnt@rev.pol.eqc.uiv.p1.generic-ppi-source.p-signflip-lt\endcsname{1}
\expandafter\def\csname pct@rev.pol.eqc.uiv.p1.generic-ppi-source.p-signflip-lt\endcsname{58.3}
\expandafter\def\csname ppt@rev.pol.eqc.uiv.p1.generic-ppi-source.p-signflip-lt\endcsname{$+$58.3}
\expandafter\def\csname val@rev.pol.eqc.uiv.p1.generic-ppi-source.score\endcsname{0.805}
\expandafter\def\csname cnt@rev.pol.eqc.uiv.p1.generic-ppi-source.score\endcsname{1}
\expandafter\def\csname pct@rev.pol.eqc.uiv.p1.generic-ppi-source.score\endcsname{80.5}
\expandafter\def\csname ppt@rev.pol.eqc.uiv.p1.generic-ppi-source.score\endcsname{$+$80.5}
\expandafter\def\csname val@rev.pol.eqc.uiv.p1.rcr-cost\endcsname{257.113}
\expandafter\def\csname cnt@rev.pol.eqc.uiv.p1.rcr-cost\endcsname{257}
\expandafter\def\csname pct@rev.pol.eqc.uiv.p1.rcr-cost\endcsname{25711.3}
\expandafter\def\csname ppt@rev.pol.eqc.uiv.p1.rcr-cost\endcsname{$+$25711.3}
\expandafter\def\csname val@rev.pol.eqc.uiv.p1.replay-ppi-uniform.cost\endcsname{257.117}
\expandafter\def\csname cnt@rev.pol.eqc.uiv.p1.replay-ppi-uniform.cost\endcsname{257}
\expandafter\def\csname pct@rev.pol.eqc.uiv.p1.replay-ppi-uniform.cost\endcsname{25711.7}
\expandafter\def\csname ppt@rev.pol.eqc.uiv.p1.replay-ppi-uniform.cost\endcsname{$+$25711.7}
\expandafter\def\csname val@rev.pol.eqc.uiv.p1.replay-ppi-uniform.lin.ci95.0\endcsname{-0.001}
\expandafter\def\csname cnt@rev.pol.eqc.uiv.p1.replay-ppi-uniform.lin.ci95.0\endcsname{0}
\expandafter\def\csname pct@rev.pol.eqc.uiv.p1.replay-ppi-uniform.lin.ci95.0\endcsname{$-$0.1}
\expandafter\def\csname ppt@rev.pol.eqc.uiv.p1.replay-ppi-uniform.lin.ci95.0\endcsname{$-$0.1}
\expandafter\def\csname val@rev.pol.eqc.uiv.p1.replay-ppi-uniform.lin.ci95.1\endcsname{0.004}
\expandafter\def\csname cnt@rev.pol.eqc.uiv.p1.replay-ppi-uniform.lin.ci95.1\endcsname{0}
\expandafter\def\csname pct@rev.pol.eqc.uiv.p1.replay-ppi-uniform.lin.ci95.1\endcsname{0.4}
\expandafter\def\csname ppt@rev.pol.eqc.uiv.p1.replay-ppi-uniform.lin.ci95.1\endcsname{$+$0.4}
\expandafter\def\csname val@rev.pol.eqc.uiv.p1.replay-ppi-uniform.lin.mean\endcsname{0.001}
\expandafter\def\csname cnt@rev.pol.eqc.uiv.p1.replay-ppi-uniform.lin.mean\endcsname{0}
\expandafter\def\csname pct@rev.pol.eqc.uiv.p1.replay-ppi-uniform.lin.mean\endcsname{0.1}
\expandafter\def\csname ppt@rev.pol.eqc.uiv.p1.replay-ppi-uniform.lin.mean\endcsname{$+$0.1}
\expandafter\def\csname val@rev.pol.eqc.uiv.p1.replay-ppi-uniform.linN.ci95.0\endcsname{0.002}
\expandafter\def\csname cnt@rev.pol.eqc.uiv.p1.replay-ppi-uniform.linN.ci95.0\endcsname{0}
\expandafter\def\csname pct@rev.pol.eqc.uiv.p1.replay-ppi-uniform.linN.ci95.0\endcsname{0.2}
\expandafter\def\csname ppt@rev.pol.eqc.uiv.p1.replay-ppi-uniform.linN.ci95.0\endcsname{$+$0.2}
\expandafter\def\csname val@rev.pol.eqc.uiv.p1.replay-ppi-uniform.linN.ci95.1\endcsname{0.007}
\expandafter\def\csname cnt@rev.pol.eqc.uiv.p1.replay-ppi-uniform.linN.ci95.1\endcsname{0}
\expandafter\def\csname pct@rev.pol.eqc.uiv.p1.replay-ppi-uniform.linN.ci95.1\endcsname{0.7}
\expandafter\def\csname ppt@rev.pol.eqc.uiv.p1.replay-ppi-uniform.linN.ci95.1\endcsname{$+$0.7}
\expandafter\def\csname val@rev.pol.eqc.uiv.p1.replay-ppi-uniform.linN.mean\endcsname{0.004}
\expandafter\def\csname cnt@rev.pol.eqc.uiv.p1.replay-ppi-uniform.linN.mean\endcsname{0}
\expandafter\def\csname pct@rev.pol.eqc.uiv.p1.replay-ppi-uniform.linN.mean\endcsname{0.4}
\expandafter\def\csname ppt@rev.pol.eqc.uiv.p1.replay-ppi-uniform.linN.mean\endcsname{$+$0.4}
\expandafter\def\csname val@rev.pol.eqc.uiv.p1.replay-ppi-uniform.minus.ci95.0\endcsname{0.002}
\expandafter\def\csname cnt@rev.pol.eqc.uiv.p1.replay-ppi-uniform.minus.ci95.0\endcsname{0}
\expandafter\def\csname pct@rev.pol.eqc.uiv.p1.replay-ppi-uniform.minus.ci95.0\endcsname{0.2}
\expandafter\def\csname ppt@rev.pol.eqc.uiv.p1.replay-ppi-uniform.minus.ci95.0\endcsname{$+$0.2}
\expandafter\def\csname val@rev.pol.eqc.uiv.p1.replay-ppi-uniform.minus.ci95.1\endcsname{0.007}
\expandafter\def\csname cnt@rev.pol.eqc.uiv.p1.replay-ppi-uniform.minus.ci95.1\endcsname{0}
\expandafter\def\csname pct@rev.pol.eqc.uiv.p1.replay-ppi-uniform.minus.ci95.1\endcsname{0.7}
\expandafter\def\csname ppt@rev.pol.eqc.uiv.p1.replay-ppi-uniform.minus.ci95.1\endcsname{$+$0.7}
\expandafter\def\csname val@rev.pol.eqc.uiv.p1.replay-ppi-uniform.minus.mean\endcsname{0.004}
\expandafter\def\csname cnt@rev.pol.eqc.uiv.p1.replay-ppi-uniform.minus.mean\endcsname{0}
\expandafter\def\csname pct@rev.pol.eqc.uiv.p1.replay-ppi-uniform.minus.mean\endcsname{0.4}
\expandafter\def\csname ppt@rev.pol.eqc.uiv.p1.replay-ppi-uniform.minus.mean\endcsname{$+$0.4}
\expandafter\def\csname val@rev.pol.eqc.uiv.p1.replay-ppi-uniform.minus.p-gt\endcsname{0.000}
\expandafter\def\csname cnt@rev.pol.eqc.uiv.p1.replay-ppi-uniform.minus.p-gt\endcsname{0}
\expandafter\def\csname pct@rev.pol.eqc.uiv.p1.replay-ppi-uniform.minus.p-gt\endcsname{0.0}
\expandafter\def\csname ppt@rev.pol.eqc.uiv.p1.replay-ppi-uniform.minus.p-gt\endcsname{0.0}
\expandafter\def\csname val@rev.pol.eqc.uiv.p1.replay-ppi-uniform.p-signflip-gt\endcsname{0.001}
\expandafter\def\csname cnt@rev.pol.eqc.uiv.p1.replay-ppi-uniform.p-signflip-gt\endcsname{0}
\expandafter\def\csname pct@rev.pol.eqc.uiv.p1.replay-ppi-uniform.p-signflip-gt\endcsname{0.1}
\expandafter\def\csname ppt@rev.pol.eqc.uiv.p1.replay-ppi-uniform.p-signflip-gt\endcsname{$+$0.1}
\expandafter\def\csname val@rev.pol.eqc.uiv.p1.replay-ppi-uniform.p-signflip-lt\endcsname{0.999}
\expandafter\def\csname cnt@rev.pol.eqc.uiv.p1.replay-ppi-uniform.p-signflip-lt\endcsname{1}
\expandafter\def\csname pct@rev.pol.eqc.uiv.p1.replay-ppi-uniform.p-signflip-lt\endcsname{99.9}
\expandafter\def\csname ppt@rev.pol.eqc.uiv.p1.replay-ppi-uniform.p-signflip-lt\endcsname{$+$99.9}
\expandafter\def\csname val@rev.pol.eqc.uiv.p1.replay-ppi-uniform.score\endcsname{0.800}
\expandafter\def\csname cnt@rev.pol.eqc.uiv.p1.replay-ppi-uniform.score\endcsname{1}
\expandafter\def\csname pct@rev.pol.eqc.uiv.p1.replay-ppi-uniform.score\endcsname{80.0}
\expandafter\def\csname ppt@rev.pol.eqc.uiv.p1.replay-ppi-uniform.score\endcsname{$+$80.0}
\expandafter\def\csname val@rev.pol.eqc.uiv.p1.uniform-fresh.cost\endcsname{257.120}
\expandafter\def\csname cnt@rev.pol.eqc.uiv.p1.uniform-fresh.cost\endcsname{257}
\expandafter\def\csname pct@rev.pol.eqc.uiv.p1.uniform-fresh.cost\endcsname{25712.0}
\expandafter\def\csname ppt@rev.pol.eqc.uiv.p1.uniform-fresh.cost\endcsname{$+$25712.0}
\expandafter\def\csname val@rev.pol.eqc.uiv.p1.uniform-fresh.lin.ci95.0\endcsname{-0.001}
\expandafter\def\csname cnt@rev.pol.eqc.uiv.p1.uniform-fresh.lin.ci95.0\endcsname{0}
\expandafter\def\csname pct@rev.pol.eqc.uiv.p1.uniform-fresh.lin.ci95.0\endcsname{$-$0.1}
\expandafter\def\csname ppt@rev.pol.eqc.uiv.p1.uniform-fresh.lin.ci95.0\endcsname{$-$0.1}
\expandafter\def\csname val@rev.pol.eqc.uiv.p1.uniform-fresh.lin.ci95.1\endcsname{0.003}
\expandafter\def\csname cnt@rev.pol.eqc.uiv.p1.uniform-fresh.lin.ci95.1\endcsname{0}
\expandafter\def\csname pct@rev.pol.eqc.uiv.p1.uniform-fresh.lin.ci95.1\endcsname{0.3}
\expandafter\def\csname ppt@rev.pol.eqc.uiv.p1.uniform-fresh.lin.ci95.1\endcsname{$+$0.3}
\expandafter\def\csname val@rev.pol.eqc.uiv.p1.uniform-fresh.lin.mean\endcsname{0.001}
\expandafter\def\csname cnt@rev.pol.eqc.uiv.p1.uniform-fresh.lin.mean\endcsname{0}
\expandafter\def\csname pct@rev.pol.eqc.uiv.p1.uniform-fresh.lin.mean\endcsname{0.1}
\expandafter\def\csname ppt@rev.pol.eqc.uiv.p1.uniform-fresh.lin.mean\endcsname{$+$0.1}
\expandafter\def\csname val@rev.pol.eqc.uiv.p1.uniform-fresh.linN.ci95.0\endcsname{0.004}
\expandafter\def\csname cnt@rev.pol.eqc.uiv.p1.uniform-fresh.linN.ci95.0\endcsname{0}
\expandafter\def\csname pct@rev.pol.eqc.uiv.p1.uniform-fresh.linN.ci95.0\endcsname{0.4}
\expandafter\def\csname ppt@rev.pol.eqc.uiv.p1.uniform-fresh.linN.ci95.0\endcsname{$+$0.4}
\expandafter\def\csname val@rev.pol.eqc.uiv.p1.uniform-fresh.linN.ci95.1\endcsname{0.014}
\expandafter\def\csname cnt@rev.pol.eqc.uiv.p1.uniform-fresh.linN.ci95.1\endcsname{0}
\expandafter\def\csname pct@rev.pol.eqc.uiv.p1.uniform-fresh.linN.ci95.1\endcsname{1.4}
\expandafter\def\csname ppt@rev.pol.eqc.uiv.p1.uniform-fresh.linN.ci95.1\endcsname{$+$1.4}
\expandafter\def\csname val@rev.pol.eqc.uiv.p1.uniform-fresh.linN.mean\endcsname{0.009}
\expandafter\def\csname cnt@rev.pol.eqc.uiv.p1.uniform-fresh.linN.mean\endcsname{0}
\expandafter\def\csname pct@rev.pol.eqc.uiv.p1.uniform-fresh.linN.mean\endcsname{0.9}
\expandafter\def\csname ppt@rev.pol.eqc.uiv.p1.uniform-fresh.linN.mean\endcsname{$+$0.9}
\expandafter\def\csname val@rev.pol.eqc.uiv.p1.uniform-fresh.minus.ci95.0\endcsname{0.003}
\expandafter\def\csname cnt@rev.pol.eqc.uiv.p1.uniform-fresh.minus.ci95.0\endcsname{0}
\expandafter\def\csname pct@rev.pol.eqc.uiv.p1.uniform-fresh.minus.ci95.0\endcsname{0.3}
\expandafter\def\csname ppt@rev.pol.eqc.uiv.p1.uniform-fresh.minus.ci95.0\endcsname{$+$0.3}
\expandafter\def\csname val@rev.pol.eqc.uiv.p1.uniform-fresh.minus.ci95.1\endcsname{0.012}
\expandafter\def\csname cnt@rev.pol.eqc.uiv.p1.uniform-fresh.minus.ci95.1\endcsname{0}
\expandafter\def\csname pct@rev.pol.eqc.uiv.p1.uniform-fresh.minus.ci95.1\endcsname{1.2}
\expandafter\def\csname ppt@rev.pol.eqc.uiv.p1.uniform-fresh.minus.ci95.1\endcsname{$+$1.2}
\expandafter\def\csname val@rev.pol.eqc.uiv.p1.uniform-fresh.minus.mean\endcsname{0.007}
\expandafter\def\csname cnt@rev.pol.eqc.uiv.p1.uniform-fresh.minus.mean\endcsname{0}
\expandafter\def\csname pct@rev.pol.eqc.uiv.p1.uniform-fresh.minus.mean\endcsname{0.7}
\expandafter\def\csname ppt@rev.pol.eqc.uiv.p1.uniform-fresh.minus.mean\endcsname{$+$0.7}
\expandafter\def\csname val@rev.pol.eqc.uiv.p1.uniform-fresh.minus.p-gt\endcsname{0.000}
\expandafter\def\csname cnt@rev.pol.eqc.uiv.p1.uniform-fresh.minus.p-gt\endcsname{0}
\expandafter\def\csname pct@rev.pol.eqc.uiv.p1.uniform-fresh.minus.p-gt\endcsname{0.0}
\expandafter\def\csname ppt@rev.pol.eqc.uiv.p1.uniform-fresh.minus.p-gt\endcsname{0.0}
\expandafter\def\csname val@rev.pol.eqc.uiv.p1.uniform-fresh.p-signflip-gt\endcsname{0.001}
\expandafter\def\csname cnt@rev.pol.eqc.uiv.p1.uniform-fresh.p-signflip-gt\endcsname{0}
\expandafter\def\csname pct@rev.pol.eqc.uiv.p1.uniform-fresh.p-signflip-gt\endcsname{0.1}
\expandafter\def\csname ppt@rev.pol.eqc.uiv.p1.uniform-fresh.p-signflip-gt\endcsname{$+$0.1}
\expandafter\def\csname val@rev.pol.eqc.uiv.p1.uniform-fresh.p-signflip-lt\endcsname{0.999}
\expandafter\def\csname cnt@rev.pol.eqc.uiv.p1.uniform-fresh.p-signflip-lt\endcsname{1}
\expandafter\def\csname pct@rev.pol.eqc.uiv.p1.uniform-fresh.p-signflip-lt\endcsname{99.9}
\expandafter\def\csname ppt@rev.pol.eqc.uiv.p1.uniform-fresh.p-signflip-lt\endcsname{$+$99.9}
\expandafter\def\csname val@rev.pol.eqc.uiv.p1.uniform-fresh.score\endcsname{0.798}
\expandafter\def\csname cnt@rev.pol.eqc.uiv.p1.uniform-fresh.score\endcsname{1}
\expandafter\def\csname pct@rev.pol.eqc.uiv.p1.uniform-fresh.score\endcsname{79.8}
\expandafter\def\csname ppt@rev.pol.eqc.uiv.p1.uniform-fresh.score\endcsname{$+$79.8}
\expandafter\def\csname val@rev.pol.eqc.uiv.p2.best-single.cost\endcsname{256.029}
\expandafter\def\csname cnt@rev.pol.eqc.uiv.p2.best-single.cost\endcsname{256}
\expandafter\def\csname pct@rev.pol.eqc.uiv.p2.best-single.cost\endcsname{25602.9}
\expandafter\def\csname ppt@rev.pol.eqc.uiv.p2.best-single.cost\endcsname{$+$25602.9}
\expandafter\def\csname val@rev.pol.eqc.uiv.p2.best-single.lin.ci95.0\endcsname{-0.009}
\expandafter\def\csname cnt@rev.pol.eqc.uiv.p2.best-single.lin.ci95.0\endcsname{0}
\expandafter\def\csname pct@rev.pol.eqc.uiv.p2.best-single.lin.ci95.0\endcsname{$-$0.9}
\expandafter\def\csname ppt@rev.pol.eqc.uiv.p2.best-single.lin.ci95.0\endcsname{$-$0.9}
\expandafter\def\csname val@rev.pol.eqc.uiv.p2.best-single.lin.ci95.1\endcsname{0.025}
\expandafter\def\csname cnt@rev.pol.eqc.uiv.p2.best-single.lin.ci95.1\endcsname{0}
\expandafter\def\csname pct@rev.pol.eqc.uiv.p2.best-single.lin.ci95.1\endcsname{2.5}
\expandafter\def\csname ppt@rev.pol.eqc.uiv.p2.best-single.lin.ci95.1\endcsname{$+$2.5}
\expandafter\def\csname val@rev.pol.eqc.uiv.p2.best-single.lin.mean\endcsname{0.007}
\expandafter\def\csname cnt@rev.pol.eqc.uiv.p2.best-single.lin.mean\endcsname{0}
\expandafter\def\csname pct@rev.pol.eqc.uiv.p2.best-single.lin.mean\endcsname{0.7}
\expandafter\def\csname ppt@rev.pol.eqc.uiv.p2.best-single.lin.mean\endcsname{$+$0.7}
\expandafter\def\csname val@rev.pol.eqc.uiv.p2.best-single.linN.ci95.0\endcsname{-0.008}
\expandafter\def\csname cnt@rev.pol.eqc.uiv.p2.best-single.linN.ci95.0\endcsname{0}
\expandafter\def\csname pct@rev.pol.eqc.uiv.p2.best-single.linN.ci95.0\endcsname{$-$0.8}
\expandafter\def\csname ppt@rev.pol.eqc.uiv.p2.best-single.linN.ci95.0\endcsname{$-$0.8}
\expandafter\def\csname val@rev.pol.eqc.uiv.p2.best-single.linN.ci95.1\endcsname{0.022}
\expandafter\def\csname cnt@rev.pol.eqc.uiv.p2.best-single.linN.ci95.1\endcsname{0}
\expandafter\def\csname pct@rev.pol.eqc.uiv.p2.best-single.linN.ci95.1\endcsname{2.2}
\expandafter\def\csname ppt@rev.pol.eqc.uiv.p2.best-single.linN.ci95.1\endcsname{$+$2.2}
\expandafter\def\csname val@rev.pol.eqc.uiv.p2.best-single.linN.mean\endcsname{0.006}
\expandafter\def\csname cnt@rev.pol.eqc.uiv.p2.best-single.linN.mean\endcsname{0}
\expandafter\def\csname pct@rev.pol.eqc.uiv.p2.best-single.linN.mean\endcsname{0.6}
\expandafter\def\csname ppt@rev.pol.eqc.uiv.p2.best-single.linN.mean\endcsname{$+$0.6}
\expandafter\def\csname val@rev.pol.eqc.uiv.p2.best-single.minus.ci95.0\endcsname{-0.005}
\expandafter\def\csname cnt@rev.pol.eqc.uiv.p2.best-single.minus.ci95.0\endcsname{0}
\expandafter\def\csname pct@rev.pol.eqc.uiv.p2.best-single.minus.ci95.0\endcsname{$-$0.5}
\expandafter\def\csname ppt@rev.pol.eqc.uiv.p2.best-single.minus.ci95.0\endcsname{$-$0.5}
\expandafter\def\csname val@rev.pol.eqc.uiv.p2.best-single.minus.ci95.1\endcsname{0.017}
\expandafter\def\csname cnt@rev.pol.eqc.uiv.p2.best-single.minus.ci95.1\endcsname{0}
\expandafter\def\csname pct@rev.pol.eqc.uiv.p2.best-single.minus.ci95.1\endcsname{1.7}
\expandafter\def\csname ppt@rev.pol.eqc.uiv.p2.best-single.minus.ci95.1\endcsname{$+$1.7}
\expandafter\def\csname val@rev.pol.eqc.uiv.p2.best-single.minus.mean\endcsname{0.005}
\expandafter\def\csname cnt@rev.pol.eqc.uiv.p2.best-single.minus.mean\endcsname{0}
\expandafter\def\csname pct@rev.pol.eqc.uiv.p2.best-single.minus.mean\endcsname{0.5}
\expandafter\def\csname ppt@rev.pol.eqc.uiv.p2.best-single.minus.mean\endcsname{$+$0.5}
\expandafter\def\csname val@rev.pol.eqc.uiv.p2.best-single.minus.p-gt\endcsname{0.186}
\expandafter\def\csname cnt@rev.pol.eqc.uiv.p2.best-single.minus.p-gt\endcsname{0}
\expandafter\def\csname pct@rev.pol.eqc.uiv.p2.best-single.minus.p-gt\endcsname{18.6}
\expandafter\def\csname ppt@rev.pol.eqc.uiv.p2.best-single.minus.p-gt\endcsname{$+$18.6}
\expandafter\def\csname val@rev.pol.eqc.uiv.p2.best-single.p-signflip-gt\endcsname{0.217}
\expandafter\def\csname cnt@rev.pol.eqc.uiv.p2.best-single.p-signflip-gt\endcsname{0}
\expandafter\def\csname pct@rev.pol.eqc.uiv.p2.best-single.p-signflip-gt\endcsname{21.7}
\expandafter\def\csname ppt@rev.pol.eqc.uiv.p2.best-single.p-signflip-gt\endcsname{$+$21.7}
\expandafter\def\csname val@rev.pol.eqc.uiv.p2.best-single.p-signflip-lt\endcsname{0.784}
\expandafter\def\csname cnt@rev.pol.eqc.uiv.p2.best-single.p-signflip-lt\endcsname{1}
\expandafter\def\csname pct@rev.pol.eqc.uiv.p2.best-single.p-signflip-lt\endcsname{78.4}
\expandafter\def\csname ppt@rev.pol.eqc.uiv.p2.best-single.p-signflip-lt\endcsname{$+$78.4}
\expandafter\def\csname val@rev.pol.eqc.uiv.p2.best-single.score\endcsname{0.824}
\expandafter\def\csname cnt@rev.pol.eqc.uiv.p2.best-single.score\endcsname{1}
\expandafter\def\csname pct@rev.pol.eqc.uiv.p2.best-single.score\endcsname{82.4}
\expandafter\def\csname ppt@rev.pol.eqc.uiv.p2.best-single.score\endcsname{$+$82.4}
\expandafter\def\csname val@rev.pol.eqc.uiv.p2.generic-ppi-source.cost\endcsname{256.029}
\expandafter\def\csname cnt@rev.pol.eqc.uiv.p2.generic-ppi-source.cost\endcsname{256}
\expandafter\def\csname pct@rev.pol.eqc.uiv.p2.generic-ppi-source.cost\endcsname{25602.9}
\expandafter\def\csname ppt@rev.pol.eqc.uiv.p2.generic-ppi-source.cost\endcsname{$+$25602.9}
\expandafter\def\csname val@rev.pol.eqc.uiv.p2.generic-ppi-source.lin.ci95.0\endcsname{-0.004}
\expandafter\def\csname cnt@rev.pol.eqc.uiv.p2.generic-ppi-source.lin.ci95.0\endcsname{0}
\expandafter\def\csname pct@rev.pol.eqc.uiv.p2.generic-ppi-source.lin.ci95.0\endcsname{$-$0.4}
\expandafter\def\csname ppt@rev.pol.eqc.uiv.p2.generic-ppi-source.lin.ci95.0\endcsname{$-$0.4}
\expandafter\def\csname val@rev.pol.eqc.uiv.p2.generic-ppi-source.lin.ci95.1\endcsname{0.008}
\expandafter\def\csname cnt@rev.pol.eqc.uiv.p2.generic-ppi-source.lin.ci95.1\endcsname{0}
\expandafter\def\csname pct@rev.pol.eqc.uiv.p2.generic-ppi-source.lin.ci95.1\endcsname{0.8}
\expandafter\def\csname ppt@rev.pol.eqc.uiv.p2.generic-ppi-source.lin.ci95.1\endcsname{$+$0.8}
\expandafter\def\csname val@rev.pol.eqc.uiv.p2.generic-ppi-source.lin.mean\endcsname{0.001}
\expandafter\def\csname cnt@rev.pol.eqc.uiv.p2.generic-ppi-source.lin.mean\endcsname{0}
\expandafter\def\csname pct@rev.pol.eqc.uiv.p2.generic-ppi-source.lin.mean\endcsname{0.1}
\expandafter\def\csname ppt@rev.pol.eqc.uiv.p2.generic-ppi-source.lin.mean\endcsname{$+$0.1}
\expandafter\def\csname val@rev.pol.eqc.uiv.p2.generic-ppi-source.linN.ci95.0\endcsname{-0.002}
\expandafter\def\csname cnt@rev.pol.eqc.uiv.p2.generic-ppi-source.linN.ci95.0\endcsname{0}
\expandafter\def\csname pct@rev.pol.eqc.uiv.p2.generic-ppi-source.linN.ci95.0\endcsname{$-$0.2}
\expandafter\def\csname ppt@rev.pol.eqc.uiv.p2.generic-ppi-source.linN.ci95.0\endcsname{$-$0.2}
\expandafter\def\csname val@rev.pol.eqc.uiv.p2.generic-ppi-source.linN.ci95.1\endcsname{0.003}
\expandafter\def\csname cnt@rev.pol.eqc.uiv.p2.generic-ppi-source.linN.ci95.1\endcsname{0}
\expandafter\def\csname pct@rev.pol.eqc.uiv.p2.generic-ppi-source.linN.ci95.1\endcsname{0.3}
\expandafter\def\csname ppt@rev.pol.eqc.uiv.p2.generic-ppi-source.linN.ci95.1\endcsname{$+$0.3}
\expandafter\def\csname val@rev.pol.eqc.uiv.p2.generic-ppi-source.linN.mean\endcsname{-0.000}
\expandafter\def\csname cnt@rev.pol.eqc.uiv.p2.generic-ppi-source.linN.mean\endcsname{0}
\expandafter\def\csname pct@rev.pol.eqc.uiv.p2.generic-ppi-source.linN.mean\endcsname{0.0}
\expandafter\def\csname ppt@rev.pol.eqc.uiv.p2.generic-ppi-source.linN.mean\endcsname{0.0}
\expandafter\def\csname val@rev.pol.eqc.uiv.p2.generic-ppi-source.minus.ci95.0\endcsname{-0.000}
\expandafter\def\csname cnt@rev.pol.eqc.uiv.p2.generic-ppi-source.minus.ci95.0\endcsname{0}
\expandafter\def\csname pct@rev.pol.eqc.uiv.p2.generic-ppi-source.minus.ci95.0\endcsname{0.0}
\expandafter\def\csname ppt@rev.pol.eqc.uiv.p2.generic-ppi-source.minus.ci95.0\endcsname{0.0}
\expandafter\def\csname val@rev.pol.eqc.uiv.p2.generic-ppi-source.minus.ci95.1\endcsname{0.004}
\expandafter\def\csname cnt@rev.pol.eqc.uiv.p2.generic-ppi-source.minus.ci95.1\endcsname{0}
\expandafter\def\csname pct@rev.pol.eqc.uiv.p2.generic-ppi-source.minus.ci95.1\endcsname{0.4}
\expandafter\def\csname ppt@rev.pol.eqc.uiv.p2.generic-ppi-source.minus.ci95.1\endcsname{$+$0.4}
\expandafter\def\csname val@rev.pol.eqc.uiv.p2.generic-ppi-source.minus.mean\endcsname{0.002}
\expandafter\def\csname cnt@rev.pol.eqc.uiv.p2.generic-ppi-source.minus.mean\endcsname{0}
\expandafter\def\csname pct@rev.pol.eqc.uiv.p2.generic-ppi-source.minus.mean\endcsname{0.2}
\expandafter\def\csname ppt@rev.pol.eqc.uiv.p2.generic-ppi-source.minus.mean\endcsname{$+$0.2}
\expandafter\def\csname val@rev.pol.eqc.uiv.p2.generic-ppi-source.minus.p-gt\endcsname{0.048}
\expandafter\def\csname cnt@rev.pol.eqc.uiv.p2.generic-ppi-source.minus.p-gt\endcsname{0}
\expandafter\def\csname pct@rev.pol.eqc.uiv.p2.generic-ppi-source.minus.p-gt\endcsname{4.8}
\expandafter\def\csname ppt@rev.pol.eqc.uiv.p2.generic-ppi-source.minus.p-gt\endcsname{$+$4.8}
\expandafter\def\csname val@rev.pol.eqc.uiv.p2.generic-ppi-source.p-signflip-gt\endcsname{0.084}
\expandafter\def\csname cnt@rev.pol.eqc.uiv.p2.generic-ppi-source.p-signflip-gt\endcsname{0}
\expandafter\def\csname pct@rev.pol.eqc.uiv.p2.generic-ppi-source.p-signflip-gt\endcsname{8.4}
\expandafter\def\csname ppt@rev.pol.eqc.uiv.p2.generic-ppi-source.p-signflip-gt\endcsname{$+$8.4}
\expandafter\def\csname val@rev.pol.eqc.uiv.p2.generic-ppi-source.p-signflip-lt\endcsname{0.917}
\expandafter\def\csname cnt@rev.pol.eqc.uiv.p2.generic-ppi-source.p-signflip-lt\endcsname{1}
\expandafter\def\csname pct@rev.pol.eqc.uiv.p2.generic-ppi-source.p-signflip-lt\endcsname{91.7}
\expandafter\def\csname ppt@rev.pol.eqc.uiv.p2.generic-ppi-source.p-signflip-lt\endcsname{$+$91.7}
\expandafter\def\csname val@rev.pol.eqc.uiv.p2.generic-ppi-source.score\endcsname{0.827}
\expandafter\def\csname cnt@rev.pol.eqc.uiv.p2.generic-ppi-source.score\endcsname{1}
\expandafter\def\csname pct@rev.pol.eqc.uiv.p2.generic-ppi-source.score\endcsname{82.7}
\expandafter\def\csname ppt@rev.pol.eqc.uiv.p2.generic-ppi-source.score\endcsname{$+$82.7}
\expandafter\def\csname val@rev.pol.eqc.uiv.p2.rcr-cost\endcsname{256.032}
\expandafter\def\csname cnt@rev.pol.eqc.uiv.p2.rcr-cost\endcsname{256}
\expandafter\def\csname pct@rev.pol.eqc.uiv.p2.rcr-cost\endcsname{25603.2}
\expandafter\def\csname ppt@rev.pol.eqc.uiv.p2.rcr-cost\endcsname{$+$25603.2}
\expandafter\def\csname val@rev.pol.eqc.uiv.p2.replay-ppi-uniform.cost\endcsname{256.031}
\expandafter\def\csname cnt@rev.pol.eqc.uiv.p2.replay-ppi-uniform.cost\endcsname{256}
\expandafter\def\csname pct@rev.pol.eqc.uiv.p2.replay-ppi-uniform.cost\endcsname{25603.1}
\expandafter\def\csname ppt@rev.pol.eqc.uiv.p2.replay-ppi-uniform.cost\endcsname{$+$25603.1}
\expandafter\def\csname val@rev.pol.eqc.uiv.p2.replay-ppi-uniform.lin.ci95.0\endcsname{0.000}
\expandafter\def\csname cnt@rev.pol.eqc.uiv.p2.replay-ppi-uniform.lin.ci95.0\endcsname{0}
\expandafter\def\csname pct@rev.pol.eqc.uiv.p2.replay-ppi-uniform.lin.ci95.0\endcsname{0.0}
\expandafter\def\csname ppt@rev.pol.eqc.uiv.p2.replay-ppi-uniform.lin.ci95.0\endcsname{0.0}
\expandafter\def\csname val@rev.pol.eqc.uiv.p2.replay-ppi-uniform.lin.ci95.1\endcsname{0.004}
\expandafter\def\csname cnt@rev.pol.eqc.uiv.p2.replay-ppi-uniform.lin.ci95.1\endcsname{0}
\expandafter\def\csname pct@rev.pol.eqc.uiv.p2.replay-ppi-uniform.lin.ci95.1\endcsname{0.4}
\expandafter\def\csname ppt@rev.pol.eqc.uiv.p2.replay-ppi-uniform.lin.ci95.1\endcsname{$+$0.4}
\expandafter\def\csname val@rev.pol.eqc.uiv.p2.replay-ppi-uniform.lin.mean\endcsname{0.002}
\expandafter\def\csname cnt@rev.pol.eqc.uiv.p2.replay-ppi-uniform.lin.mean\endcsname{0}
\expandafter\def\csname pct@rev.pol.eqc.uiv.p2.replay-ppi-uniform.lin.mean\endcsname{0.2}
\expandafter\def\csname ppt@rev.pol.eqc.uiv.p2.replay-ppi-uniform.lin.mean\endcsname{$+$0.2}
\expandafter\def\csname val@rev.pol.eqc.uiv.p2.replay-ppi-uniform.linN.ci95.0\endcsname{0.000}
\expandafter\def\csname cnt@rev.pol.eqc.uiv.p2.replay-ppi-uniform.linN.ci95.0\endcsname{0}
\expandafter\def\csname pct@rev.pol.eqc.uiv.p2.replay-ppi-uniform.linN.ci95.0\endcsname{0.0}
\expandafter\def\csname ppt@rev.pol.eqc.uiv.p2.replay-ppi-uniform.linN.ci95.0\endcsname{0.0}
\expandafter\def\csname val@rev.pol.eqc.uiv.p2.replay-ppi-uniform.linN.ci95.1\endcsname{0.002}
\expandafter\def\csname cnt@rev.pol.eqc.uiv.p2.replay-ppi-uniform.linN.ci95.1\endcsname{0}
\expandafter\def\csname pct@rev.pol.eqc.uiv.p2.replay-ppi-uniform.linN.ci95.1\endcsname{0.2}
\expandafter\def\csname ppt@rev.pol.eqc.uiv.p2.replay-ppi-uniform.linN.ci95.1\endcsname{$+$0.2}
\expandafter\def\csname val@rev.pol.eqc.uiv.p2.replay-ppi-uniform.linN.mean\endcsname{0.001}
\expandafter\def\csname cnt@rev.pol.eqc.uiv.p2.replay-ppi-uniform.linN.mean\endcsname{0}
\expandafter\def\csname pct@rev.pol.eqc.uiv.p2.replay-ppi-uniform.linN.mean\endcsname{0.1}
\expandafter\def\csname ppt@rev.pol.eqc.uiv.p2.replay-ppi-uniform.linN.mean\endcsname{$+$0.1}
\expandafter\def\csname val@rev.pol.eqc.uiv.p2.replay-ppi-uniform.minus.ci95.0\endcsname{0.000}
\expandafter\def\csname cnt@rev.pol.eqc.uiv.p2.replay-ppi-uniform.minus.ci95.0\endcsname{0}
\expandafter\def\csname pct@rev.pol.eqc.uiv.p2.replay-ppi-uniform.minus.ci95.0\endcsname{0.0}
\expandafter\def\csname ppt@rev.pol.eqc.uiv.p2.replay-ppi-uniform.minus.ci95.0\endcsname{0.0}
\expandafter\def\csname val@rev.pol.eqc.uiv.p2.replay-ppi-uniform.minus.ci95.1\endcsname{0.002}
\expandafter\def\csname cnt@rev.pol.eqc.uiv.p2.replay-ppi-uniform.minus.ci95.1\endcsname{0}
\expandafter\def\csname pct@rev.pol.eqc.uiv.p2.replay-ppi-uniform.minus.ci95.1\endcsname{0.2}
\expandafter\def\csname ppt@rev.pol.eqc.uiv.p2.replay-ppi-uniform.minus.ci95.1\endcsname{$+$0.2}
\expandafter\def\csname val@rev.pol.eqc.uiv.p2.replay-ppi-uniform.minus.mean\endcsname{0.001}
\expandafter\def\csname cnt@rev.pol.eqc.uiv.p2.replay-ppi-uniform.minus.mean\endcsname{0}
\expandafter\def\csname pct@rev.pol.eqc.uiv.p2.replay-ppi-uniform.minus.mean\endcsname{0.1}
\expandafter\def\csname ppt@rev.pol.eqc.uiv.p2.replay-ppi-uniform.minus.mean\endcsname{$+$0.1}
\expandafter\def\csname val@rev.pol.eqc.uiv.p2.replay-ppi-uniform.minus.p-gt\endcsname{0.003}
\expandafter\def\csname cnt@rev.pol.eqc.uiv.p2.replay-ppi-uniform.minus.p-gt\endcsname{0}
\expandafter\def\csname pct@rev.pol.eqc.uiv.p2.replay-ppi-uniform.minus.p-gt\endcsname{0.3}
\expandafter\def\csname ppt@rev.pol.eqc.uiv.p2.replay-ppi-uniform.minus.p-gt\endcsname{$+$0.3}
\expandafter\def\csname val@rev.pol.eqc.uiv.p2.replay-ppi-uniform.p-signflip-gt\endcsname{0.008}
\expandafter\def\csname cnt@rev.pol.eqc.uiv.p2.replay-ppi-uniform.p-signflip-gt\endcsname{0}
\expandafter\def\csname pct@rev.pol.eqc.uiv.p2.replay-ppi-uniform.p-signflip-gt\endcsname{0.8}
\expandafter\def\csname ppt@rev.pol.eqc.uiv.p2.replay-ppi-uniform.p-signflip-gt\endcsname{$+$0.8}
\expandafter\def\csname val@rev.pol.eqc.uiv.p2.replay-ppi-uniform.p-signflip-lt\endcsname{0.993}
\expandafter\def\csname cnt@rev.pol.eqc.uiv.p2.replay-ppi-uniform.p-signflip-lt\endcsname{1}
\expandafter\def\csname pct@rev.pol.eqc.uiv.p2.replay-ppi-uniform.p-signflip-lt\endcsname{99.3}
\expandafter\def\csname ppt@rev.pol.eqc.uiv.p2.replay-ppi-uniform.p-signflip-lt\endcsname{$+$99.3}
\expandafter\def\csname val@rev.pol.eqc.uiv.p2.replay-ppi-uniform.score\endcsname{0.827}
\expandafter\def\csname cnt@rev.pol.eqc.uiv.p2.replay-ppi-uniform.score\endcsname{1}
\expandafter\def\csname pct@rev.pol.eqc.uiv.p2.replay-ppi-uniform.score\endcsname{82.7}
\expandafter\def\csname ppt@rev.pol.eqc.uiv.p2.replay-ppi-uniform.score\endcsname{$+$82.7}
\expandafter\def\csname val@rev.pol.eqc.uiv.p2.uniform-fresh.cost\endcsname{256.029}
\expandafter\def\csname cnt@rev.pol.eqc.uiv.p2.uniform-fresh.cost\endcsname{256}
\expandafter\def\csname pct@rev.pol.eqc.uiv.p2.uniform-fresh.cost\endcsname{25602.9}
\expandafter\def\csname ppt@rev.pol.eqc.uiv.p2.uniform-fresh.cost\endcsname{$+$25602.9}
\expandafter\def\csname val@rev.pol.eqc.uiv.p2.uniform-fresh.lin.ci95.0\endcsname{-0.005}
\expandafter\def\csname cnt@rev.pol.eqc.uiv.p2.uniform-fresh.lin.ci95.0\endcsname{0}
\expandafter\def\csname pct@rev.pol.eqc.uiv.p2.uniform-fresh.lin.ci95.0\endcsname{$-$0.5}
\expandafter\def\csname ppt@rev.pol.eqc.uiv.p2.uniform-fresh.lin.ci95.0\endcsname{$-$0.5}
\expandafter\def\csname val@rev.pol.eqc.uiv.p2.uniform-fresh.lin.ci95.1\endcsname{0.005}
\expandafter\def\csname cnt@rev.pol.eqc.uiv.p2.uniform-fresh.lin.ci95.1\endcsname{0}
\expandafter\def\csname pct@rev.pol.eqc.uiv.p2.uniform-fresh.lin.ci95.1\endcsname{0.5}
\expandafter\def\csname ppt@rev.pol.eqc.uiv.p2.uniform-fresh.lin.ci95.1\endcsname{$+$0.5}
\expandafter\def\csname val@rev.pol.eqc.uiv.p2.uniform-fresh.lin.mean\endcsname{-0.000}
\expandafter\def\csname cnt@rev.pol.eqc.uiv.p2.uniform-fresh.lin.mean\endcsname{0}
\expandafter\def\csname pct@rev.pol.eqc.uiv.p2.uniform-fresh.lin.mean\endcsname{0.0}
\expandafter\def\csname ppt@rev.pol.eqc.uiv.p2.uniform-fresh.lin.mean\endcsname{0.0}
\expandafter\def\csname val@rev.pol.eqc.uiv.p2.uniform-fresh.linN.ci95.0\endcsname{-0.002}
\expandafter\def\csname cnt@rev.pol.eqc.uiv.p2.uniform-fresh.linN.ci95.0\endcsname{0}
\expandafter\def\csname pct@rev.pol.eqc.uiv.p2.uniform-fresh.linN.ci95.0\endcsname{$-$0.2}
\expandafter\def\csname ppt@rev.pol.eqc.uiv.p2.uniform-fresh.linN.ci95.0\endcsname{$-$0.2}
\expandafter\def\csname val@rev.pol.eqc.uiv.p2.uniform-fresh.linN.ci95.1\endcsname{0.006}
\expandafter\def\csname cnt@rev.pol.eqc.uiv.p2.uniform-fresh.linN.ci95.1\endcsname{0}
\expandafter\def\csname pct@rev.pol.eqc.uiv.p2.uniform-fresh.linN.ci95.1\endcsname{0.6}
\expandafter\def\csname ppt@rev.pol.eqc.uiv.p2.uniform-fresh.linN.ci95.1\endcsname{$+$0.6}
\expandafter\def\csname val@rev.pol.eqc.uiv.p2.uniform-fresh.linN.mean\endcsname{0.002}
\expandafter\def\csname cnt@rev.pol.eqc.uiv.p2.uniform-fresh.linN.mean\endcsname{0}
\expandafter\def\csname pct@rev.pol.eqc.uiv.p2.uniform-fresh.linN.mean\endcsname{0.2}
\expandafter\def\csname ppt@rev.pol.eqc.uiv.p2.uniform-fresh.linN.mean\endcsname{$+$0.2}
\expandafter\def\csname val@rev.pol.eqc.uiv.p2.uniform-fresh.minus.ci95.0\endcsname{-0.001}
\expandafter\def\csname cnt@rev.pol.eqc.uiv.p2.uniform-fresh.minus.ci95.0\endcsname{0}
\expandafter\def\csname pct@rev.pol.eqc.uiv.p2.uniform-fresh.minus.ci95.0\endcsname{$-$0.1}
\expandafter\def\csname ppt@rev.pol.eqc.uiv.p2.uniform-fresh.minus.ci95.0\endcsname{$-$0.1}
\expandafter\def\csname val@rev.pol.eqc.uiv.p2.uniform-fresh.minus.ci95.1\endcsname{0.010}
\expandafter\def\csname cnt@rev.pol.eqc.uiv.p2.uniform-fresh.minus.ci95.1\endcsname{0}
\expandafter\def\csname pct@rev.pol.eqc.uiv.p2.uniform-fresh.minus.ci95.1\endcsname{1.0}
\expandafter\def\csname ppt@rev.pol.eqc.uiv.p2.uniform-fresh.minus.ci95.1\endcsname{$+$1.0}
\expandafter\def\csname val@rev.pol.eqc.uiv.p2.uniform-fresh.minus.mean\endcsname{0.004}
\expandafter\def\csname cnt@rev.pol.eqc.uiv.p2.uniform-fresh.minus.mean\endcsname{0}
\expandafter\def\csname pct@rev.pol.eqc.uiv.p2.uniform-fresh.minus.mean\endcsname{0.4}
\expandafter\def\csname ppt@rev.pol.eqc.uiv.p2.uniform-fresh.minus.mean\endcsname{$+$0.4}
\expandafter\def\csname val@rev.pol.eqc.uiv.p2.uniform-fresh.minus.p-gt\endcsname{0.079}
\expandafter\def\csname cnt@rev.pol.eqc.uiv.p2.uniform-fresh.minus.p-gt\endcsname{0}
\expandafter\def\csname pct@rev.pol.eqc.uiv.p2.uniform-fresh.minus.p-gt\endcsname{7.9}
\expandafter\def\csname ppt@rev.pol.eqc.uiv.p2.uniform-fresh.minus.p-gt\endcsname{$+$7.9}
\expandafter\def\csname val@rev.pol.eqc.uiv.p2.uniform-fresh.p-signflip-gt\endcsname{0.105}
\expandafter\def\csname cnt@rev.pol.eqc.uiv.p2.uniform-fresh.p-signflip-gt\endcsname{0}
\expandafter\def\csname pct@rev.pol.eqc.uiv.p2.uniform-fresh.p-signflip-gt\endcsname{10.5}
\expandafter\def\csname ppt@rev.pol.eqc.uiv.p2.uniform-fresh.p-signflip-gt\endcsname{$+$10.5}
\expandafter\def\csname val@rev.pol.eqc.uiv.p2.uniform-fresh.p-signflip-lt\endcsname{0.896}
\expandafter\def\csname cnt@rev.pol.eqc.uiv.p2.uniform-fresh.p-signflip-lt\endcsname{1}
\expandafter\def\csname pct@rev.pol.eqc.uiv.p2.uniform-fresh.p-signflip-lt\endcsname{89.6}
\expandafter\def\csname ppt@rev.pol.eqc.uiv.p2.uniform-fresh.p-signflip-lt\endcsname{$+$89.6}
\expandafter\def\csname val@rev.pol.eqc.uiv.p2.uniform-fresh.score\endcsname{0.825}
\expandafter\def\csname cnt@rev.pol.eqc.uiv.p2.uniform-fresh.score\endcsname{1}
\expandafter\def\csname pct@rev.pol.eqc.uiv.p2.uniform-fresh.score\endcsname{82.5}
\expandafter\def\csname ppt@rev.pol.eqc.uiv.p2.uniform-fresh.score\endcsname{$+$82.5}
\expandafter\def\csname val@rev.pol.eqc.uniform-fresh.max\endcsname{0.007}
\expandafter\def\csname cnt@rev.pol.eqc.uniform-fresh.max\endcsname{0}
\expandafter\def\csname pct@rev.pol.eqc.uniform-fresh.max\endcsname{0.7}
\expandafter\def\csname ppt@rev.pol.eqc.uniform-fresh.max\endcsname{$+$0.7}
\expandafter\def\csname val@rev.pol.eqc.uniform-fresh.min\endcsname{-0.011}
\expandafter\def\csname cnt@rev.pol.eqc.uniform-fresh.min\endcsname{0}
\expandafter\def\csname pct@rev.pol.eqc.uniform-fresh.min\endcsname{$-$1.1}
\expandafter\def\csname ppt@rev.pol.eqc.uniform-fresh.min\endcsname{$-$1.1}
\expandafter\def\csname val@rev.pol.eqc.uniform-fresh.n-ci-above-0\endcsname{1}
\expandafter\def\csname cnt@rev.pol.eqc.uniform-fresh.n-ci-above-0\endcsname{1}
\expandafter\def\csname pct@rev.pol.eqc.uniform-fresh.n-ci-above-0\endcsname{100.0}
\expandafter\def\csname ppt@rev.pol.eqc.uniform-fresh.n-ci-above-0\endcsname{$+$100.0}
\expandafter\def\csname val@rev.pol.eqc.uniform-fresh.n-ci-above-0-lin20\endcsname{0}
\expandafter\def\csname cnt@rev.pol.eqc.uniform-fresh.n-ci-above-0-lin20\endcsname{0}
\expandafter\def\csname pct@rev.pol.eqc.uniform-fresh.n-ci-above-0-lin20\endcsname{0.0}
\expandafter\def\csname ppt@rev.pol.eqc.uniform-fresh.n-ci-above-0-lin20\endcsname{0.0}
\expandafter\def\csname val@rev.pol.eqc.uniform-fresh.n-ci-above-0-linN\endcsname{1}
\expandafter\def\csname cnt@rev.pol.eqc.uniform-fresh.n-ci-above-0-linN\endcsname{1}
\expandafter\def\csname pct@rev.pol.eqc.uniform-fresh.n-ci-above-0-linN\endcsname{100.0}
\expandafter\def\csname ppt@rev.pol.eqc.uniform-fresh.n-ci-above-0-linN\endcsname{$+$100.0}
\expandafter\def\csname val@rev.pol.eqc.uniform-fresh.n-ci-below-0\endcsname{2}
\expandafter\def\csname cnt@rev.pol.eqc.uniform-fresh.n-ci-below-0\endcsname{2}
\expandafter\def\csname pct@rev.pol.eqc.uniform-fresh.n-ci-below-0\endcsname{200.0}
\expandafter\def\csname ppt@rev.pol.eqc.uniform-fresh.n-ci-below-0\endcsname{$+$200.0}
\expandafter\def\csname val@rev.pol.eqc.uniform-fresh.n-ci-below-0-lin20\endcsname{2}
\expandafter\def\csname cnt@rev.pol.eqc.uniform-fresh.n-ci-below-0-lin20\endcsname{2}
\expandafter\def\csname pct@rev.pol.eqc.uniform-fresh.n-ci-below-0-lin20\endcsname{200.0}
\expandafter\def\csname ppt@rev.pol.eqc.uniform-fresh.n-ci-below-0-lin20\endcsname{$+$200.0}
\expandafter\def\csname val@rev.pol.eqc.uniform-fresh.n-ci-below-0-linN\endcsname{2}
\expandafter\def\csname cnt@rev.pol.eqc.uniform-fresh.n-ci-below-0-linN\endcsname{2}
\expandafter\def\csname pct@rev.pol.eqc.uniform-fresh.n-ci-below-0-linN\endcsname{200.0}
\expandafter\def\csname ppt@rev.pol.eqc.uniform-fresh.n-ci-below-0-linN\endcsname{$+$200.0}
\expandafter\def\csname val@rev.pol.eqc.uniform-fresh.n-sign-change-vs-lin\endcsname{1}
\expandafter\def\csname cnt@rev.pol.eqc.uniform-fresh.n-sign-change-vs-lin\endcsname{1}
\expandafter\def\csname pct@rev.pol.eqc.uniform-fresh.n-sign-change-vs-lin\endcsname{100.0}
\expandafter\def\csname ppt@rev.pol.eqc.uniform-fresh.n-sign-change-vs-lin\endcsname{$+$100.0}
\expandafter\def\csname val@rev.pol.holm.f20.n-rejected\endcsname{7}
\expandafter\def\csname cnt@rev.pol.holm.f20.n-rejected\endcsname{7}
\expandafter\def\csname pct@rev.pol.holm.f20.n-rejected\endcsname{700.0}
\expandafter\def\csname ppt@rev.pol.holm.f20.n-rejected\endcsname{$+$700.0}
\expandafter\def\csname val@rev.pol.holm.f20.per-setting.n-rejected\endcsname{7}
\expandafter\def\csname cnt@rev.pol.holm.f20.per-setting.n-rejected\endcsname{7}
\expandafter\def\csname pct@rev.pol.holm.f20.per-setting.n-rejected\endcsname{700.0}
\expandafter\def\csname ppt@rev.pol.holm.f20.per-setting.n-rejected\endcsname{$+$700.0}
\expandafter\def\csname val@rev.pol.holm.family-size\endcsname{8}
\expandafter\def\csname cnt@rev.pol.holm.family-size\endcsname{8}
\expandafter\def\csname pct@rev.pol.holm.family-size\endcsname{800.0}
\expandafter\def\csname ppt@rev.pol.holm.family-size\endcsname{$+$800.0}
\expandafter\def\csname val@rev.pol.holm.mc.max-abs-z\endcsname{1.439}
\expandafter\def\csname cnt@rev.pol.holm.mc.max-abs-z\endcsname{1}
\expandafter\def\csname pct@rev.pol.holm.mc.max-abs-z\endcsname{143.9}
\expandafter\def\csname ppt@rev.pol.holm.mc.max-abs-z\endcsname{$+$143.9}
\expandafter\def\csname val@rev.pol.holm.mc.max-se20\endcsname{0.007}
\expandafter\def\csname cnt@rev.pol.holm.mc.max-se20\endcsname{0}
\expandafter\def\csname pct@rev.pol.holm.mc.max-se20\endcsname{0.7}
\expandafter\def\csname ppt@rev.pol.holm.mc.max-se20\endcsname{$+$0.7}
\expandafter\def\csname val@rev.pol.holm.mc.min-se20\endcsname{0.003}
\expandafter\def\csname cnt@rev.pol.holm.mc.min-se20\endcsname{0}
\expandafter\def\csname pct@rev.pol.holm.mc.min-se20\endcsname{0.3}
\expandafter\def\csname ppt@rev.pol.holm.mc.min-se20\endcsname{$+$0.3}
\expandafter\def\csname val@rev.pol.holm.mc.sspro.p1.rcr-vs-generic-ppi-source.sd\endcsname{0.023}
\expandafter\def\csname cnt@rev.pol.holm.mc.sspro.p1.rcr-vs-generic-ppi-source.sd\endcsname{0}
\expandafter\def\csname pct@rev.pol.holm.mc.sspro.p1.rcr-vs-generic-ppi-source.sd\endcsname{2.3}
\expandafter\def\csname ppt@rev.pol.holm.mc.sspro.p1.rcr-vs-generic-ppi-source.sd\endcsname{$+$2.3}
\expandafter\def\csname val@rev.pol.holm.mc.sspro.p1.rcr-vs-generic-ppi-source.se20\endcsname{0.005}
\expandafter\def\csname cnt@rev.pol.holm.mc.sspro.p1.rcr-vs-generic-ppi-source.se20\endcsname{0}
\expandafter\def\csname pct@rev.pol.holm.mc.sspro.p1.rcr-vs-generic-ppi-source.se20\endcsname{0.5}
\expandafter\def\csname ppt@rev.pol.holm.mc.sspro.p1.rcr-vs-generic-ppi-source.se20\endcsname{$+$0.5}
\expandafter\def\csname val@rev.pol.holm.mc.sspro.p1.rcr-vs-generic-ppi-source.seN\endcsname{0.000}
\expandafter\def\csname cnt@rev.pol.holm.mc.sspro.p1.rcr-vs-generic-ppi-source.seN\endcsname{0}
\expandafter\def\csname pct@rev.pol.holm.mc.sspro.p1.rcr-vs-generic-ppi-source.seN\endcsname{0.0}
\expandafter\def\csname ppt@rev.pol.holm.mc.sspro.p1.rcr-vs-generic-ppi-source.seN\endcsname{0.0}
\expandafter\def\csname val@rev.pol.holm.mc.sspro.p1.rcr-vs-generic-ppi-source.share-neg\endcsname{0.087}
\expandafter\def\csname cnt@rev.pol.holm.mc.sspro.p1.rcr-vs-generic-ppi-source.share-neg\endcsname{0}
\expandafter\def\csname pct@rev.pol.holm.mc.sspro.p1.rcr-vs-generic-ppi-source.share-neg\endcsname{8.7}
\expandafter\def\csname ppt@rev.pol.holm.mc.sspro.p1.rcr-vs-generic-ppi-source.share-neg\endcsname{$+$8.7}
\expandafter\def\csname val@rev.pol.holm.mc.sspro.p1.rcr-vs-generic-ppi-source.share-pos\endcsname{0.287}
\expandafter\def\csname cnt@rev.pol.holm.mc.sspro.p1.rcr-vs-generic-ppi-source.share-pos\endcsname{0}
\expandafter\def\csname pct@rev.pol.holm.mc.sspro.p1.rcr-vs-generic-ppi-source.share-pos\endcsname{28.7}
\expandafter\def\csname ppt@rev.pol.holm.mc.sspro.p1.rcr-vs-generic-ppi-source.share-pos\endcsname{$+$28.7}
\expandafter\def\csname val@rev.pol.holm.mc.sspro.p1.rcr-vs-generic-ppi-source.z\endcsname{-0.449}
\expandafter\def\csname cnt@rev.pol.holm.mc.sspro.p1.rcr-vs-generic-ppi-source.z\endcsname{0}
\expandafter\def\csname pct@rev.pol.holm.mc.sspro.p1.rcr-vs-generic-ppi-source.z\endcsname{$-$44.9}
\expandafter\def\csname ppt@rev.pol.holm.mc.sspro.p1.rcr-vs-generic-ppi-source.z\endcsname{$-$44.9}
\expandafter\def\csname val@rev.pol.holm.mc.sspro.p1.rcr-vs-replay-ppi-uniform.sd\endcsname{0.019}
\expandafter\def\csname cnt@rev.pol.holm.mc.sspro.p1.rcr-vs-replay-ppi-uniform.sd\endcsname{0}
\expandafter\def\csname pct@rev.pol.holm.mc.sspro.p1.rcr-vs-replay-ppi-uniform.sd\endcsname{1.9}
\expandafter\def\csname ppt@rev.pol.holm.mc.sspro.p1.rcr-vs-replay-ppi-uniform.sd\endcsname{$+$1.9}
\expandafter\def\csname val@rev.pol.holm.mc.sspro.p1.rcr-vs-replay-ppi-uniform.se20\endcsname{0.004}
\expandafter\def\csname cnt@rev.pol.holm.mc.sspro.p1.rcr-vs-replay-ppi-uniform.se20\endcsname{0}
\expandafter\def\csname pct@rev.pol.holm.mc.sspro.p1.rcr-vs-replay-ppi-uniform.se20\endcsname{0.4}
\expandafter\def\csname ppt@rev.pol.holm.mc.sspro.p1.rcr-vs-replay-ppi-uniform.se20\endcsname{$+$0.4}
\expandafter\def\csname val@rev.pol.holm.mc.sspro.p1.rcr-vs-replay-ppi-uniform.seN\endcsname{0.000}
\expandafter\def\csname cnt@rev.pol.holm.mc.sspro.p1.rcr-vs-replay-ppi-uniform.seN\endcsname{0}
\expandafter\def\csname pct@rev.pol.holm.mc.sspro.p1.rcr-vs-replay-ppi-uniform.seN\endcsname{0.0}
\expandafter\def\csname ppt@rev.pol.holm.mc.sspro.p1.rcr-vs-replay-ppi-uniform.seN\endcsname{0.0}
\expandafter\def\csname val@rev.pol.holm.mc.sspro.p1.rcr-vs-replay-ppi-uniform.share-neg\endcsname{0.092}
\expandafter\def\csname cnt@rev.pol.holm.mc.sspro.p1.rcr-vs-replay-ppi-uniform.share-neg\endcsname{0}
\expandafter\def\csname pct@rev.pol.holm.mc.sspro.p1.rcr-vs-replay-ppi-uniform.share-neg\endcsname{9.2}
\expandafter\def\csname ppt@rev.pol.holm.mc.sspro.p1.rcr-vs-replay-ppi-uniform.share-neg\endcsname{$+$9.2}
\expandafter\def\csname val@rev.pol.holm.mc.sspro.p1.rcr-vs-replay-ppi-uniform.share-pos\endcsname{0.178}
\expandafter\def\csname cnt@rev.pol.holm.mc.sspro.p1.rcr-vs-replay-ppi-uniform.share-pos\endcsname{0}
\expandafter\def\csname pct@rev.pol.holm.mc.sspro.p1.rcr-vs-replay-ppi-uniform.share-pos\endcsname{17.8}
\expandafter\def\csname ppt@rev.pol.holm.mc.sspro.p1.rcr-vs-replay-ppi-uniform.share-pos\endcsname{$+$17.8}
\expandafter\def\csname val@rev.pol.holm.mc.sspro.p1.rcr-vs-replay-ppi-uniform.z\endcsname{1.334}
\expandafter\def\csname cnt@rev.pol.holm.mc.sspro.p1.rcr-vs-replay-ppi-uniform.z\endcsname{1}
\expandafter\def\csname pct@rev.pol.holm.mc.sspro.p1.rcr-vs-replay-ppi-uniform.z\endcsname{133.4}
\expandafter\def\csname ppt@rev.pol.holm.mc.sspro.p1.rcr-vs-replay-ppi-uniform.z\endcsname{$+$133.4}
\expandafter\def\csname val@rev.pol.holm.mc.sspro.p2.rcr-vs-generic-ppi-source.sd\endcsname{0.030}
\expandafter\def\csname cnt@rev.pol.holm.mc.sspro.p2.rcr-vs-generic-ppi-source.sd\endcsname{0}
\expandafter\def\csname pct@rev.pol.holm.mc.sspro.p2.rcr-vs-generic-ppi-source.sd\endcsname{3.0}
\expandafter\def\csname ppt@rev.pol.holm.mc.sspro.p2.rcr-vs-generic-ppi-source.sd\endcsname{$+$3.0}
\expandafter\def\csname val@rev.pol.holm.mc.sspro.p2.rcr-vs-generic-ppi-source.se20\endcsname{0.007}
\expandafter\def\csname cnt@rev.pol.holm.mc.sspro.p2.rcr-vs-generic-ppi-source.se20\endcsname{0}
\expandafter\def\csname pct@rev.pol.holm.mc.sspro.p2.rcr-vs-generic-ppi-source.se20\endcsname{0.7}
\expandafter\def\csname ppt@rev.pol.holm.mc.sspro.p2.rcr-vs-generic-ppi-source.se20\endcsname{$+$0.7}
\expandafter\def\csname val@rev.pol.holm.mc.sspro.p2.rcr-vs-generic-ppi-source.seN\endcsname{0.000}
\expandafter\def\csname cnt@rev.pol.holm.mc.sspro.p2.rcr-vs-generic-ppi-source.seN\endcsname{0}
\expandafter\def\csname pct@rev.pol.holm.mc.sspro.p2.rcr-vs-generic-ppi-source.seN\endcsname{0.0}
\expandafter\def\csname ppt@rev.pol.holm.mc.sspro.p2.rcr-vs-generic-ppi-source.seN\endcsname{0.0}
\expandafter\def\csname val@rev.pol.holm.mc.sspro.p2.rcr-vs-generic-ppi-source.share-neg\endcsname{0.262}
\expandafter\def\csname cnt@rev.pol.holm.mc.sspro.p2.rcr-vs-generic-ppi-source.share-neg\endcsname{0}
\expandafter\def\csname pct@rev.pol.holm.mc.sspro.p2.rcr-vs-generic-ppi-source.share-neg\endcsname{26.2}
\expandafter\def\csname ppt@rev.pol.holm.mc.sspro.p2.rcr-vs-generic-ppi-source.share-neg\endcsname{$+$26.2}
\expandafter\def\csname val@rev.pol.holm.mc.sspro.p2.rcr-vs-generic-ppi-source.share-pos\endcsname{0.275}
\expandafter\def\csname cnt@rev.pol.holm.mc.sspro.p2.rcr-vs-generic-ppi-source.share-pos\endcsname{0}
\expandafter\def\csname pct@rev.pol.holm.mc.sspro.p2.rcr-vs-generic-ppi-source.share-pos\endcsname{27.5}
\expandafter\def\csname ppt@rev.pol.holm.mc.sspro.p2.rcr-vs-generic-ppi-source.share-pos\endcsname{$+$27.5}
\expandafter\def\csname val@rev.pol.holm.mc.sspro.p2.rcr-vs-generic-ppi-source.z\endcsname{1.439}
\expandafter\def\csname cnt@rev.pol.holm.mc.sspro.p2.rcr-vs-generic-ppi-source.z\endcsname{1}
\expandafter\def\csname pct@rev.pol.holm.mc.sspro.p2.rcr-vs-generic-ppi-source.z\endcsname{143.9}
\expandafter\def\csname ppt@rev.pol.holm.mc.sspro.p2.rcr-vs-generic-ppi-source.z\endcsname{$+$143.9}
\expandafter\def\csname val@rev.pol.holm.mc.sspro.p2.rcr-vs-replay-ppi-uniform.sd\endcsname{0.030}
\expandafter\def\csname cnt@rev.pol.holm.mc.sspro.p2.rcr-vs-replay-ppi-uniform.sd\endcsname{0}
\expandafter\def\csname pct@rev.pol.holm.mc.sspro.p2.rcr-vs-replay-ppi-uniform.sd\endcsname{3.0}
\expandafter\def\csname ppt@rev.pol.holm.mc.sspro.p2.rcr-vs-replay-ppi-uniform.sd\endcsname{$+$3.0}
\expandafter\def\csname val@rev.pol.holm.mc.sspro.p2.rcr-vs-replay-ppi-uniform.se20\endcsname{0.007}
\expandafter\def\csname cnt@rev.pol.holm.mc.sspro.p2.rcr-vs-replay-ppi-uniform.se20\endcsname{0}
\expandafter\def\csname pct@rev.pol.holm.mc.sspro.p2.rcr-vs-replay-ppi-uniform.se20\endcsname{0.7}
\expandafter\def\csname ppt@rev.pol.holm.mc.sspro.p2.rcr-vs-replay-ppi-uniform.se20\endcsname{$+$0.7}
\expandafter\def\csname val@rev.pol.holm.mc.sspro.p2.rcr-vs-replay-ppi-uniform.seN\endcsname{0.000}
\expandafter\def\csname cnt@rev.pol.holm.mc.sspro.p2.rcr-vs-replay-ppi-uniform.seN\endcsname{0}
\expandafter\def\csname pct@rev.pol.holm.mc.sspro.p2.rcr-vs-replay-ppi-uniform.seN\endcsname{0.0}
\expandafter\def\csname ppt@rev.pol.holm.mc.sspro.p2.rcr-vs-replay-ppi-uniform.seN\endcsname{0.0}
\expandafter\def\csname val@rev.pol.holm.mc.sspro.p2.rcr-vs-replay-ppi-uniform.share-neg\endcsname{0.261}
\expandafter\def\csname cnt@rev.pol.holm.mc.sspro.p2.rcr-vs-replay-ppi-uniform.share-neg\endcsname{0}
\expandafter\def\csname pct@rev.pol.holm.mc.sspro.p2.rcr-vs-replay-ppi-uniform.share-neg\endcsname{26.1}
\expandafter\def\csname ppt@rev.pol.holm.mc.sspro.p2.rcr-vs-replay-ppi-uniform.share-neg\endcsname{$+$26.1}
\expandafter\def\csname val@rev.pol.holm.mc.sspro.p2.rcr-vs-replay-ppi-uniform.share-pos\endcsname{0.276}
\expandafter\def\csname cnt@rev.pol.holm.mc.sspro.p2.rcr-vs-replay-ppi-uniform.share-pos\endcsname{0}
\expandafter\def\csname pct@rev.pol.holm.mc.sspro.p2.rcr-vs-replay-ppi-uniform.share-pos\endcsname{27.6}
\expandafter\def\csname ppt@rev.pol.holm.mc.sspro.p2.rcr-vs-replay-ppi-uniform.share-pos\endcsname{$+$27.6}
\expandafter\def\csname val@rev.pol.holm.mc.sspro.p2.rcr-vs-replay-ppi-uniform.z\endcsname{1.417}
\expandafter\def\csname cnt@rev.pol.holm.mc.sspro.p2.rcr-vs-replay-ppi-uniform.z\endcsname{1}
\expandafter\def\csname pct@rev.pol.holm.mc.sspro.p2.rcr-vs-replay-ppi-uniform.z\endcsname{141.7}
\expandafter\def\csname ppt@rev.pol.holm.mc.sspro.p2.rcr-vs-replay-ppi-uniform.z\endcsname{$+$141.7}
\expandafter\def\csname val@rev.pol.holm.mc.uiv.p1.rcr-vs-generic-ppi-source.sd\endcsname{0.030}
\expandafter\def\csname cnt@rev.pol.holm.mc.uiv.p1.rcr-vs-generic-ppi-source.sd\endcsname{0}
\expandafter\def\csname pct@rev.pol.holm.mc.uiv.p1.rcr-vs-generic-ppi-source.sd\endcsname{3.0}
\expandafter\def\csname ppt@rev.pol.holm.mc.uiv.p1.rcr-vs-generic-ppi-source.sd\endcsname{$+$3.0}
\expandafter\def\csname val@rev.pol.holm.mc.uiv.p1.rcr-vs-generic-ppi-source.se20\endcsname{0.007}
\expandafter\def\csname cnt@rev.pol.holm.mc.uiv.p1.rcr-vs-generic-ppi-source.se20\endcsname{0}
\expandafter\def\csname pct@rev.pol.holm.mc.uiv.p1.rcr-vs-generic-ppi-source.se20\endcsname{0.7}
\expandafter\def\csname ppt@rev.pol.holm.mc.uiv.p1.rcr-vs-generic-ppi-source.se20\endcsname{$+$0.7}
\expandafter\def\csname val@rev.pol.holm.mc.uiv.p1.rcr-vs-generic-ppi-source.seN\endcsname{0.000}
\expandafter\def\csname cnt@rev.pol.holm.mc.uiv.p1.rcr-vs-generic-ppi-source.seN\endcsname{0}
\expandafter\def\csname pct@rev.pol.holm.mc.uiv.p1.rcr-vs-generic-ppi-source.seN\endcsname{0.0}
\expandafter\def\csname ppt@rev.pol.holm.mc.uiv.p1.rcr-vs-generic-ppi-source.seN\endcsname{0.0}
\expandafter\def\csname val@rev.pol.holm.mc.uiv.p1.rcr-vs-generic-ppi-source.share-neg\endcsname{0.349}
\expandafter\def\csname cnt@rev.pol.holm.mc.uiv.p1.rcr-vs-generic-ppi-source.share-neg\endcsname{0}
\expandafter\def\csname pct@rev.pol.holm.mc.uiv.p1.rcr-vs-generic-ppi-source.share-neg\endcsname{34.9}
\expandafter\def\csname ppt@rev.pol.holm.mc.uiv.p1.rcr-vs-generic-ppi-source.share-neg\endcsname{$+$34.9}
\expandafter\def\csname val@rev.pol.holm.mc.uiv.p1.rcr-vs-generic-ppi-source.share-pos\endcsname{0.551}
\expandafter\def\csname cnt@rev.pol.holm.mc.uiv.p1.rcr-vs-generic-ppi-source.share-pos\endcsname{1}
\expandafter\def\csname pct@rev.pol.holm.mc.uiv.p1.rcr-vs-generic-ppi-source.share-pos\endcsname{55.1}
\expandafter\def\csname ppt@rev.pol.holm.mc.uiv.p1.rcr-vs-generic-ppi-source.share-pos\endcsname{$+$55.1}
\expandafter\def\csname val@rev.pol.holm.mc.uiv.p1.rcr-vs-generic-ppi-source.z\endcsname{0.815}
\expandafter\def\csname cnt@rev.pol.holm.mc.uiv.p1.rcr-vs-generic-ppi-source.z\endcsname{1}
\expandafter\def\csname pct@rev.pol.holm.mc.uiv.p1.rcr-vs-generic-ppi-source.z\endcsname{81.5}
\expandafter\def\csname ppt@rev.pol.holm.mc.uiv.p1.rcr-vs-generic-ppi-source.z\endcsname{$+$81.5}
\expandafter\def\csname val@rev.pol.holm.mc.uiv.p1.rcr-vs-replay-ppi-uniform.sd\endcsname{0.028}
\expandafter\def\csname cnt@rev.pol.holm.mc.uiv.p1.rcr-vs-replay-ppi-uniform.sd\endcsname{0}
\expandafter\def\csname pct@rev.pol.holm.mc.uiv.p1.rcr-vs-replay-ppi-uniform.sd\endcsname{2.8}
\expandafter\def\csname ppt@rev.pol.holm.mc.uiv.p1.rcr-vs-replay-ppi-uniform.sd\endcsname{$+$2.8}
\expandafter\def\csname val@rev.pol.holm.mc.uiv.p1.rcr-vs-replay-ppi-uniform.se20\endcsname{0.006}
\expandafter\def\csname cnt@rev.pol.holm.mc.uiv.p1.rcr-vs-replay-ppi-uniform.se20\endcsname{0}
\expandafter\def\csname pct@rev.pol.holm.mc.uiv.p1.rcr-vs-replay-ppi-uniform.se20\endcsname{0.6}
\expandafter\def\csname ppt@rev.pol.holm.mc.uiv.p1.rcr-vs-replay-ppi-uniform.se20\endcsname{$+$0.6}
\expandafter\def\csname val@rev.pol.holm.mc.uiv.p1.rcr-vs-replay-ppi-uniform.seN\endcsname{0.000}
\expandafter\def\csname cnt@rev.pol.holm.mc.uiv.p1.rcr-vs-replay-ppi-uniform.seN\endcsname{0}
\expandafter\def\csname pct@rev.pol.holm.mc.uiv.p1.rcr-vs-replay-ppi-uniform.seN\endcsname{0.0}
\expandafter\def\csname ppt@rev.pol.holm.mc.uiv.p1.rcr-vs-replay-ppi-uniform.seN\endcsname{0.0}
\expandafter\def\csname val@rev.pol.holm.mc.uiv.p1.rcr-vs-replay-ppi-uniform.share-neg\endcsname{0.386}
\expandafter\def\csname cnt@rev.pol.holm.mc.uiv.p1.rcr-vs-replay-ppi-uniform.share-neg\endcsname{0}
\expandafter\def\csname pct@rev.pol.holm.mc.uiv.p1.rcr-vs-replay-ppi-uniform.share-neg\endcsname{38.6}
\expandafter\def\csname ppt@rev.pol.holm.mc.uiv.p1.rcr-vs-replay-ppi-uniform.share-neg\endcsname{$+$38.6}
\expandafter\def\csname val@rev.pol.holm.mc.uiv.p1.rcr-vs-replay-ppi-uniform.share-pos\endcsname{0.505}
\expandafter\def\csname cnt@rev.pol.holm.mc.uiv.p1.rcr-vs-replay-ppi-uniform.share-pos\endcsname{1}
\expandafter\def\csname pct@rev.pol.holm.mc.uiv.p1.rcr-vs-replay-ppi-uniform.share-pos\endcsname{50.5}
\expandafter\def\csname ppt@rev.pol.holm.mc.uiv.p1.rcr-vs-replay-ppi-uniform.share-pos\endcsname{$+$50.5}
\expandafter\def\csname val@rev.pol.holm.mc.uiv.p1.rcr-vs-replay-ppi-uniform.z\endcsname{-0.461}
\expandafter\def\csname cnt@rev.pol.holm.mc.uiv.p1.rcr-vs-replay-ppi-uniform.z\endcsname{0}
\expandafter\def\csname pct@rev.pol.holm.mc.uiv.p1.rcr-vs-replay-ppi-uniform.z\endcsname{$-$46.1}
\expandafter\def\csname ppt@rev.pol.holm.mc.uiv.p1.rcr-vs-replay-ppi-uniform.z\endcsname{$-$46.1}
\expandafter\def\csname val@rev.pol.holm.mc.uiv.p2.rcr-vs-generic-ppi-source.sd\endcsname{0.012}
\expandafter\def\csname cnt@rev.pol.holm.mc.uiv.p2.rcr-vs-generic-ppi-source.sd\endcsname{0}
\expandafter\def\csname pct@rev.pol.holm.mc.uiv.p2.rcr-vs-generic-ppi-source.sd\endcsname{1.2}
\expandafter\def\csname ppt@rev.pol.holm.mc.uiv.p2.rcr-vs-generic-ppi-source.sd\endcsname{$+$1.2}
\expandafter\def\csname val@rev.pol.holm.mc.uiv.p2.rcr-vs-generic-ppi-source.se20\endcsname{0.003}
\expandafter\def\csname cnt@rev.pol.holm.mc.uiv.p2.rcr-vs-generic-ppi-source.se20\endcsname{0}
\expandafter\def\csname pct@rev.pol.holm.mc.uiv.p2.rcr-vs-generic-ppi-source.se20\endcsname{0.3}
\expandafter\def\csname ppt@rev.pol.holm.mc.uiv.p2.rcr-vs-generic-ppi-source.se20\endcsname{$+$0.3}
\expandafter\def\csname val@rev.pol.holm.mc.uiv.p2.rcr-vs-generic-ppi-source.seN\endcsname{0.000}
\expandafter\def\csname cnt@rev.pol.holm.mc.uiv.p2.rcr-vs-generic-ppi-source.seN\endcsname{0}
\expandafter\def\csname pct@rev.pol.holm.mc.uiv.p2.rcr-vs-generic-ppi-source.seN\endcsname{0.0}
\expandafter\def\csname ppt@rev.pol.holm.mc.uiv.p2.rcr-vs-generic-ppi-source.seN\endcsname{0.0}
\expandafter\def\csname val@rev.pol.holm.mc.uiv.p2.rcr-vs-generic-ppi-source.share-neg\endcsname{0.392}
\expandafter\def\csname cnt@rev.pol.holm.mc.uiv.p2.rcr-vs-generic-ppi-source.share-neg\endcsname{0}
\expandafter\def\csname pct@rev.pol.holm.mc.uiv.p2.rcr-vs-generic-ppi-source.share-neg\endcsname{39.2}
\expandafter\def\csname ppt@rev.pol.holm.mc.uiv.p2.rcr-vs-generic-ppi-source.share-neg\endcsname{$+$39.2}
\expandafter\def\csname val@rev.pol.holm.mc.uiv.p2.rcr-vs-generic-ppi-source.share-pos\endcsname{0.465}
\expandafter\def\csname cnt@rev.pol.holm.mc.uiv.p2.rcr-vs-generic-ppi-source.share-pos\endcsname{0}
\expandafter\def\csname pct@rev.pol.holm.mc.uiv.p2.rcr-vs-generic-ppi-source.share-pos\endcsname{46.5}
\expandafter\def\csname ppt@rev.pol.holm.mc.uiv.p2.rcr-vs-generic-ppi-source.share-pos\endcsname{$+$46.5}
\expandafter\def\csname val@rev.pol.holm.mc.uiv.p2.rcr-vs-generic-ppi-source.z\endcsname{0.846}
\expandafter\def\csname cnt@rev.pol.holm.mc.uiv.p2.rcr-vs-generic-ppi-source.z\endcsname{1}
\expandafter\def\csname pct@rev.pol.holm.mc.uiv.p2.rcr-vs-generic-ppi-source.z\endcsname{84.6}
\expandafter\def\csname ppt@rev.pol.holm.mc.uiv.p2.rcr-vs-generic-ppi-source.z\endcsname{$+$84.6}
\expandafter\def\csname val@rev.pol.holm.mc.uiv.p2.rcr-vs-replay-ppi-uniform.sd\endcsname{0.012}
\expandafter\def\csname cnt@rev.pol.holm.mc.uiv.p2.rcr-vs-replay-ppi-uniform.sd\endcsname{0}
\expandafter\def\csname pct@rev.pol.holm.mc.uiv.p2.rcr-vs-replay-ppi-uniform.sd\endcsname{1.2}
\expandafter\def\csname ppt@rev.pol.holm.mc.uiv.p2.rcr-vs-replay-ppi-uniform.sd\endcsname{$+$1.2}
\expandafter\def\csname val@rev.pol.holm.mc.uiv.p2.rcr-vs-replay-ppi-uniform.se20\endcsname{0.003}
\expandafter\def\csname cnt@rev.pol.holm.mc.uiv.p2.rcr-vs-replay-ppi-uniform.se20\endcsname{0}
\expandafter\def\csname pct@rev.pol.holm.mc.uiv.p2.rcr-vs-replay-ppi-uniform.se20\endcsname{0.3}
\expandafter\def\csname ppt@rev.pol.holm.mc.uiv.p2.rcr-vs-replay-ppi-uniform.se20\endcsname{$+$0.3}
\expandafter\def\csname val@rev.pol.holm.mc.uiv.p2.rcr-vs-replay-ppi-uniform.seN\endcsname{0.000}
\expandafter\def\csname cnt@rev.pol.holm.mc.uiv.p2.rcr-vs-replay-ppi-uniform.seN\endcsname{0}
\expandafter\def\csname pct@rev.pol.holm.mc.uiv.p2.rcr-vs-replay-ppi-uniform.seN\endcsname{0.0}
\expandafter\def\csname ppt@rev.pol.holm.mc.uiv.p2.rcr-vs-replay-ppi-uniform.seN\endcsname{0.0}
\expandafter\def\csname val@rev.pol.holm.mc.uiv.p2.rcr-vs-replay-ppi-uniform.share-neg\endcsname{0.389}
\expandafter\def\csname cnt@rev.pol.holm.mc.uiv.p2.rcr-vs-replay-ppi-uniform.share-neg\endcsname{0}
\expandafter\def\csname pct@rev.pol.holm.mc.uiv.p2.rcr-vs-replay-ppi-uniform.share-neg\endcsname{38.9}
\expandafter\def\csname ppt@rev.pol.holm.mc.uiv.p2.rcr-vs-replay-ppi-uniform.share-neg\endcsname{$+$38.9}
\expandafter\def\csname val@rev.pol.holm.mc.uiv.p2.rcr-vs-replay-ppi-uniform.share-pos\endcsname{0.452}
\expandafter\def\csname cnt@rev.pol.holm.mc.uiv.p2.rcr-vs-replay-ppi-uniform.share-pos\endcsname{0}
\expandafter\def\csname pct@rev.pol.holm.mc.uiv.p2.rcr-vs-replay-ppi-uniform.share-pos\endcsname{45.2}
\expandafter\def\csname ppt@rev.pol.holm.mc.uiv.p2.rcr-vs-replay-ppi-uniform.share-pos\endcsname{$+$45.2}
\expandafter\def\csname val@rev.pol.holm.mc.uiv.p2.rcr-vs-replay-ppi-uniform.z\endcsname{0.308}
\expandafter\def\csname cnt@rev.pol.holm.mc.uiv.p2.rcr-vs-replay-ppi-uniform.z\endcsname{0}
\expandafter\def\csname pct@rev.pol.holm.mc.uiv.p2.rcr-vs-replay-ppi-uniform.z\endcsname{30.8}
\expandafter\def\csname ppt@rev.pol.holm.mc.uiv.p2.rcr-vs-replay-ppi-uniform.z\endcsname{$+$30.8}
\expandafter\def\csname val@rev.pol.holm.n-lost-rejections\endcsname{3}
\expandafter\def\csname cnt@rev.pol.holm.n-lost-rejections\endcsname{3}
\expandafter\def\csname pct@rev.pol.holm.n-lost-rejections\endcsname{300.0}
\expandafter\def\csname ppt@rev.pol.holm.n-lost-rejections\endcsname{$+$300.0}
\expandafter\def\csname val@rev.pol.holm.n-new-rejections\endcsname{1}
\expandafter\def\csname cnt@rev.pol.holm.n-new-rejections\endcsname{1}
\expandafter\def\csname pct@rev.pol.holm.n-new-rejections\endcsname{100.0}
\expandafter\def\csname ppt@rev.pol.holm.n-new-rejections\endcsname{$+$100.0}
\expandafter\def\csname val@rev.pol.holm.n-rejected-both\endcsname{4}
\expandafter\def\csname cnt@rev.pol.holm.n-rejected-both\endcsname{4}
\expandafter\def\csname pct@rev.pol.holm.n-rejected-both\endcsname{400.0}
\expandafter\def\csname ppt@rev.pol.holm.n-rejected-both\endcsname{$+$400.0}
\expandafter\def\csname val@rev.pol.holm.n.max-mean\endcsname{0.008}
\expandafter\def\csname cnt@rev.pol.holm.n.max-mean\endcsname{0}
\expandafter\def\csname pct@rev.pol.holm.n.max-mean\endcsname{0.8}
\expandafter\def\csname ppt@rev.pol.holm.n.max-mean\endcsname{$+$0.8}
\expandafter\def\csname val@rev.pol.holm.n.max-p-holm-rejected\endcsname{0.035}
\expandafter\def\csname cnt@rev.pol.holm.n.max-p-holm-rejected\endcsname{0}
\expandafter\def\csname pct@rev.pol.holm.n.max-p-holm-rejected\endcsname{3.5}
\expandafter\def\csname ppt@rev.pol.holm.n.max-p-holm-rejected\endcsname{$+$3.5}
\expandafter\def\csname val@rev.pol.holm.n.min-mean\endcsname{-0.000}
\expandafter\def\csname cnt@rev.pol.holm.n.min-mean\endcsname{0}
\expandafter\def\csname pct@rev.pol.holm.n.min-mean\endcsname{0.0}
\expandafter\def\csname ppt@rev.pol.holm.n.min-mean\endcsname{0.0}
\expandafter\def\csname val@rev.pol.holm.n.min-p-holm-notrejected\endcsname{0.137}
\expandafter\def\csname cnt@rev.pol.holm.n.min-p-holm-notrejected\endcsname{0}
\expandafter\def\csname pct@rev.pol.holm.n.min-p-holm-notrejected\endcsname{13.7}
\expandafter\def\csname ppt@rev.pol.holm.n.min-p-holm-notrejected\endcsname{$+$13.7}
\expandafter\def\csname val@rev.pol.holm.n.n-negative-mean\endcsname{1}
\expandafter\def\csname cnt@rev.pol.holm.n.n-negative-mean\endcsname{1}
\expandafter\def\csname pct@rev.pol.holm.n.n-negative-mean\endcsname{100.0}
\expandafter\def\csname ppt@rev.pol.holm.n.n-negative-mean\endcsname{$+$100.0}
\expandafter\def\csname val@rev.pol.holm.n.n-rejected\endcsname{5}
\expandafter\def\csname cnt@rev.pol.holm.n.n-rejected\endcsname{5}
\expandafter\def\csname pct@rev.pol.holm.n.n-rejected\endcsname{500.0}
\expandafter\def\csname ppt@rev.pol.holm.n.n-rejected\endcsname{$+$500.0}
\expandafter\def\csname val@rev.pol.holm.n.per-setting.n-rejected\endcsname{6}
\expandafter\def\csname cnt@rev.pol.holm.n.per-setting.n-rejected\endcsname{6}
\expandafter\def\csname pct@rev.pol.holm.n.per-setting.n-rejected\endcsname{600.0}
\expandafter\def\csname ppt@rev.pol.holm.n.per-setting.n-rejected\endcsname{$+$600.0}
\expandafter\def\csname val@rev.pol.holm.sspro.p1.rcr-vs-generic-ppi-source.f20.lb\endcsname{0.002}
\expandafter\def\csname cnt@rev.pol.holm.sspro.p1.rcr-vs-generic-ppi-source.f20.lb\endcsname{0}
\expandafter\def\csname pct@rev.pol.holm.sspro.p1.rcr-vs-generic-ppi-source.f20.lb\endcsname{0.2}
\expandafter\def\csname ppt@rev.pol.holm.sspro.p1.rcr-vs-generic-ppi-source.f20.lb\endcsname{$+$0.2}
\expandafter\def\csname val@rev.pol.holm.sspro.p1.rcr-vs-generic-ppi-source.f20.mean\endcsname{0.006}
\expandafter\def\csname cnt@rev.pol.holm.sspro.p1.rcr-vs-generic-ppi-source.f20.mean\endcsname{0}
\expandafter\def\csname pct@rev.pol.holm.sspro.p1.rcr-vs-generic-ppi-source.f20.mean\endcsname{0.6}
\expandafter\def\csname ppt@rev.pol.holm.sspro.p1.rcr-vs-generic-ppi-source.f20.mean\endcsname{$+$0.6}
\expandafter\def\csname val@rev.pol.holm.sspro.p1.rcr-vs-generic-ppi-source.f20.p-holm\endcsname{0.021}
\expandafter\def\csname cnt@rev.pol.holm.sspro.p1.rcr-vs-generic-ppi-source.f20.p-holm\endcsname{0}
\expandafter\def\csname pct@rev.pol.holm.sspro.p1.rcr-vs-generic-ppi-source.f20.p-holm\endcsname{2.1}
\expandafter\def\csname ppt@rev.pol.holm.sspro.p1.rcr-vs-generic-ppi-source.f20.p-holm\endcsname{$+$2.1}
\expandafter\def\csname val@rev.pol.holm.sspro.p1.rcr-vs-generic-ppi-source.f20.p-holm-text\endcsname{0.0212}
\expandafter\def\csname val@rev.pol.holm.sspro.p1.rcr-vs-generic-ppi-source.f20.p-raw\endcsname{0.004}
\expandafter\def\csname cnt@rev.pol.holm.sspro.p1.rcr-vs-generic-ppi-source.f20.p-raw\endcsname{0}
\expandafter\def\csname pct@rev.pol.holm.sspro.p1.rcr-vs-generic-ppi-source.f20.p-raw\endcsname{0.4}
\expandafter\def\csname ppt@rev.pol.holm.sspro.p1.rcr-vs-generic-ppi-source.f20.p-raw\endcsname{$+$0.4}
\expandafter\def\csname val@rev.pol.holm.sspro.p1.rcr-vs-generic-ppi-source.f20.reject\endcsname{1}
\expandafter\def\csname cnt@rev.pol.holm.sspro.p1.rcr-vs-generic-ppi-source.f20.reject\endcsname{1}
\expandafter\def\csname pct@rev.pol.holm.sspro.p1.rcr-vs-generic-ppi-source.f20.reject\endcsname{100.0}
\expandafter\def\csname ppt@rev.pol.holm.sspro.p1.rcr-vs-generic-ppi-source.f20.reject\endcsname{$+$100.0}
\expandafter\def\csname val@rev.pol.holm.sspro.p1.rcr-vs-generic-ppi-source.n.lb\endcsname{0.002}
\expandafter\def\csname cnt@rev.pol.holm.sspro.p1.rcr-vs-generic-ppi-source.n.lb\endcsname{0}
\expandafter\def\csname pct@rev.pol.holm.sspro.p1.rcr-vs-generic-ppi-source.n.lb\endcsname{0.2}
\expandafter\def\csname ppt@rev.pol.holm.sspro.p1.rcr-vs-generic-ppi-source.n.lb\endcsname{$+$0.2}
\expandafter\def\csname val@rev.pol.holm.sspro.p1.rcr-vs-generic-ppi-source.n.mean\endcsname{0.008}
\expandafter\def\csname cnt@rev.pol.holm.sspro.p1.rcr-vs-generic-ppi-source.n.mean\endcsname{0}
\expandafter\def\csname pct@rev.pol.holm.sspro.p1.rcr-vs-generic-ppi-source.n.mean\endcsname{0.8}
\expandafter\def\csname ppt@rev.pol.holm.sspro.p1.rcr-vs-generic-ppi-source.n.mean\endcsname{$+$0.8}
\expandafter\def\csname val@rev.pol.holm.sspro.p1.rcr-vs-generic-ppi-source.n.p-holm\endcsname{0.024}
\expandafter\def\csname cnt@rev.pol.holm.sspro.p1.rcr-vs-generic-ppi-source.n.p-holm\endcsname{0}
\expandafter\def\csname pct@rev.pol.holm.sspro.p1.rcr-vs-generic-ppi-source.n.p-holm\endcsname{2.4}
\expandafter\def\csname ppt@rev.pol.holm.sspro.p1.rcr-vs-generic-ppi-source.n.p-holm\endcsname{$+$2.4}
\expandafter\def\csname val@rev.pol.holm.sspro.p1.rcr-vs-generic-ppi-source.n.p-holm-text\endcsname{0.0237}
\expandafter\def\csname val@rev.pol.holm.sspro.p1.rcr-vs-generic-ppi-source.n.p-raw\endcsname{0.005}
\expandafter\def\csname cnt@rev.pol.holm.sspro.p1.rcr-vs-generic-ppi-source.n.p-raw\endcsname{0}
\expandafter\def\csname pct@rev.pol.holm.sspro.p1.rcr-vs-generic-ppi-source.n.p-raw\endcsname{0.5}
\expandafter\def\csname ppt@rev.pol.holm.sspro.p1.rcr-vs-generic-ppi-source.n.p-raw\endcsname{$+$0.5}
\expandafter\def\csname val@rev.pol.holm.sspro.p1.rcr-vs-generic-ppi-source.n.reject\endcsname{1}
\expandafter\def\csname cnt@rev.pol.holm.sspro.p1.rcr-vs-generic-ppi-source.n.reject\endcsname{1}
\expandafter\def\csname pct@rev.pol.holm.sspro.p1.rcr-vs-generic-ppi-source.n.reject\endcsname{100.0}
\expandafter\def\csname ppt@rev.pol.holm.sspro.p1.rcr-vs-generic-ppi-source.n.reject\endcsname{$+$100.0}
\expandafter\def\csname val@rev.pol.holm.sspro.p1.rcr-vs-replay-ppi-uniform.f20.lb\endcsname{0.002}
\expandafter\def\csname cnt@rev.pol.holm.sspro.p1.rcr-vs-replay-ppi-uniform.f20.lb\endcsname{0}
\expandafter\def\csname pct@rev.pol.holm.sspro.p1.rcr-vs-replay-ppi-uniform.f20.lb\endcsname{0.2}
\expandafter\def\csname ppt@rev.pol.holm.sspro.p1.rcr-vs-replay-ppi-uniform.f20.lb\endcsname{$+$0.2}
\expandafter\def\csname val@rev.pol.holm.sspro.p1.rcr-vs-replay-ppi-uniform.f20.mean\endcsname{0.009}
\expandafter\def\csname cnt@rev.pol.holm.sspro.p1.rcr-vs-replay-ppi-uniform.f20.mean\endcsname{0}
\expandafter\def\csname pct@rev.pol.holm.sspro.p1.rcr-vs-replay-ppi-uniform.f20.mean\endcsname{0.9}
\expandafter\def\csname ppt@rev.pol.holm.sspro.p1.rcr-vs-replay-ppi-uniform.f20.mean\endcsname{$+$0.9}
\expandafter\def\csname val@rev.pol.holm.sspro.p1.rcr-vs-replay-ppi-uniform.f20.p-holm\endcsname{0.049}
\expandafter\def\csname cnt@rev.pol.holm.sspro.p1.rcr-vs-replay-ppi-uniform.f20.p-holm\endcsname{0}
\expandafter\def\csname pct@rev.pol.holm.sspro.p1.rcr-vs-replay-ppi-uniform.f20.p-holm\endcsname{4.9}
\expandafter\def\csname ppt@rev.pol.holm.sspro.p1.rcr-vs-replay-ppi-uniform.f20.p-holm\endcsname{$+$4.9}
\expandafter\def\csname val@rev.pol.holm.sspro.p1.rcr-vs-replay-ppi-uniform.f20.p-holm-text\endcsname{0.0490}
\expandafter\def\csname val@rev.pol.holm.sspro.p1.rcr-vs-replay-ppi-uniform.f20.p-raw\endcsname{0.014}
\expandafter\def\csname cnt@rev.pol.holm.sspro.p1.rcr-vs-replay-ppi-uniform.f20.p-raw\endcsname{0}
\expandafter\def\csname pct@rev.pol.holm.sspro.p1.rcr-vs-replay-ppi-uniform.f20.p-raw\endcsname{1.4}
\expandafter\def\csname ppt@rev.pol.holm.sspro.p1.rcr-vs-replay-ppi-uniform.f20.p-raw\endcsname{$+$1.4}
\expandafter\def\csname val@rev.pol.holm.sspro.p1.rcr-vs-replay-ppi-uniform.f20.reject\endcsname{1}
\expandafter\def\csname cnt@rev.pol.holm.sspro.p1.rcr-vs-replay-ppi-uniform.f20.reject\endcsname{1}
\expandafter\def\csname pct@rev.pol.holm.sspro.p1.rcr-vs-replay-ppi-uniform.f20.reject\endcsname{100.0}
\expandafter\def\csname ppt@rev.pol.holm.sspro.p1.rcr-vs-replay-ppi-uniform.f20.reject\endcsname{$+$100.0}
\expandafter\def\csname val@rev.pol.holm.sspro.p1.rcr-vs-replay-ppi-uniform.n.lb\endcsname{0.001}
\expandafter\def\csname cnt@rev.pol.holm.sspro.p1.rcr-vs-replay-ppi-uniform.n.lb\endcsname{0}
\expandafter\def\csname pct@rev.pol.holm.sspro.p1.rcr-vs-replay-ppi-uniform.n.lb\endcsname{0.1}
\expandafter\def\csname ppt@rev.pol.holm.sspro.p1.rcr-vs-replay-ppi-uniform.n.lb\endcsname{$+$0.1}
\expandafter\def\csname val@rev.pol.holm.sspro.p1.rcr-vs-replay-ppi-uniform.n.mean\endcsname{0.003}
\expandafter\def\csname cnt@rev.pol.holm.sspro.p1.rcr-vs-replay-ppi-uniform.n.mean\endcsname{0}
\expandafter\def\csname pct@rev.pol.holm.sspro.p1.rcr-vs-replay-ppi-uniform.n.mean\endcsname{0.3}
\expandafter\def\csname ppt@rev.pol.holm.sspro.p1.rcr-vs-replay-ppi-uniform.n.mean\endcsname{$+$0.3}
\expandafter\def\csname val@rev.pol.holm.sspro.p1.rcr-vs-replay-ppi-uniform.n.p-holm\endcsname{0.035}
\expandafter\def\csname cnt@rev.pol.holm.sspro.p1.rcr-vs-replay-ppi-uniform.n.p-holm\endcsname{0}
\expandafter\def\csname pct@rev.pol.holm.sspro.p1.rcr-vs-replay-ppi-uniform.n.p-holm\endcsname{3.5}
\expandafter\def\csname ppt@rev.pol.holm.sspro.p1.rcr-vs-replay-ppi-uniform.n.p-holm\endcsname{$+$3.5}
\expandafter\def\csname val@rev.pol.holm.sspro.p1.rcr-vs-replay-ppi-uniform.n.p-holm-text\endcsname{0.0350}
\expandafter\def\csname val@rev.pol.holm.sspro.p1.rcr-vs-replay-ppi-uniform.n.p-raw\endcsname{0.009}
\expandafter\def\csname cnt@rev.pol.holm.sspro.p1.rcr-vs-replay-ppi-uniform.n.p-raw\endcsname{0}
\expandafter\def\csname pct@rev.pol.holm.sspro.p1.rcr-vs-replay-ppi-uniform.n.p-raw\endcsname{0.9}
\expandafter\def\csname ppt@rev.pol.holm.sspro.p1.rcr-vs-replay-ppi-uniform.n.p-raw\endcsname{$+$0.9}
\expandafter\def\csname val@rev.pol.holm.sspro.p1.rcr-vs-replay-ppi-uniform.n.reject\endcsname{1}
\expandafter\def\csname cnt@rev.pol.holm.sspro.p1.rcr-vs-replay-ppi-uniform.n.reject\endcsname{1}
\expandafter\def\csname pct@rev.pol.holm.sspro.p1.rcr-vs-replay-ppi-uniform.n.reject\endcsname{100.0}
\expandafter\def\csname ppt@rev.pol.holm.sspro.p1.rcr-vs-replay-ppi-uniform.n.reject\endcsname{$+$100.0}
\expandafter\def\csname val@rev.pol.holm.sspro.p2.rcr-vs-generic-ppi-source.f20.lb\endcsname{0.006}
\expandafter\def\csname cnt@rev.pol.holm.sspro.p2.rcr-vs-generic-ppi-source.f20.lb\endcsname{0}
\expandafter\def\csname pct@rev.pol.holm.sspro.p2.rcr-vs-generic-ppi-source.f20.lb\endcsname{0.6}
\expandafter\def\csname ppt@rev.pol.holm.sspro.p2.rcr-vs-generic-ppi-source.f20.lb\endcsname{$+$0.6}
\expandafter\def\csname val@rev.pol.holm.sspro.p2.rcr-vs-generic-ppi-source.f20.mean\endcsname{0.010}
\expandafter\def\csname cnt@rev.pol.holm.sspro.p2.rcr-vs-generic-ppi-source.f20.mean\endcsname{0}
\expandafter\def\csname pct@rev.pol.holm.sspro.p2.rcr-vs-generic-ppi-source.f20.mean\endcsname{1.0}
\expandafter\def\csname ppt@rev.pol.holm.sspro.p2.rcr-vs-generic-ppi-source.f20.mean\endcsname{$+$1.0}
\expandafter\def\csname val@rev.pol.holm.sspro.p2.rcr-vs-generic-ppi-source.f20.p-holm\endcsname{0.002}
\expandafter\def\csname cnt@rev.pol.holm.sspro.p2.rcr-vs-generic-ppi-source.f20.p-holm\endcsname{0}
\expandafter\def\csname pct@rev.pol.holm.sspro.p2.rcr-vs-generic-ppi-source.f20.p-holm\endcsname{0.2}
\expandafter\def\csname ppt@rev.pol.holm.sspro.p2.rcr-vs-generic-ppi-source.f20.p-holm\endcsname{$+$0.2}
\expandafter\def\csname val@rev.pol.holm.sspro.p2.rcr-vs-generic-ppi-source.f20.p-holm-text\endcsname{0.0020}
\expandafter\def\csname val@rev.pol.holm.sspro.p2.rcr-vs-generic-ppi-source.f20.p-raw\endcsname{0.000}
\expandafter\def\csname cnt@rev.pol.holm.sspro.p2.rcr-vs-generic-ppi-source.f20.p-raw\endcsname{0}
\expandafter\def\csname pct@rev.pol.holm.sspro.p2.rcr-vs-generic-ppi-source.f20.p-raw\endcsname{0.0}
\expandafter\def\csname ppt@rev.pol.holm.sspro.p2.rcr-vs-generic-ppi-source.f20.p-raw\endcsname{0.0}
\expandafter\def\csname val@rev.pol.holm.sspro.p2.rcr-vs-generic-ppi-source.f20.reject\endcsname{1}
\expandafter\def\csname cnt@rev.pol.holm.sspro.p2.rcr-vs-generic-ppi-source.f20.reject\endcsname{1}
\expandafter\def\csname pct@rev.pol.holm.sspro.p2.rcr-vs-generic-ppi-source.f20.reject\endcsname{100.0}
\expandafter\def\csname ppt@rev.pol.holm.sspro.p2.rcr-vs-generic-ppi-source.f20.reject\endcsname{$+$100.0}
\expandafter\def\csname val@rev.pol.holm.sspro.p2.rcr-vs-generic-ppi-source.n.lb\endcsname{-0.001}
\expandafter\def\csname cnt@rev.pol.holm.sspro.p2.rcr-vs-generic-ppi-source.n.lb\endcsname{0}
\expandafter\def\csname pct@rev.pol.holm.sspro.p2.rcr-vs-generic-ppi-source.n.lb\endcsname{$-$0.1}
\expandafter\def\csname ppt@rev.pol.holm.sspro.p2.rcr-vs-generic-ppi-source.n.lb\endcsname{$-$0.1}
\expandafter\def\csname val@rev.pol.holm.sspro.p2.rcr-vs-generic-ppi-source.n.mean\endcsname{-0.000}
\expandafter\def\csname cnt@rev.pol.holm.sspro.p2.rcr-vs-generic-ppi-source.n.mean\endcsname{0}
\expandafter\def\csname pct@rev.pol.holm.sspro.p2.rcr-vs-generic-ppi-source.n.mean\endcsname{0.0}
\expandafter\def\csname ppt@rev.pol.holm.sspro.p2.rcr-vs-generic-ppi-source.n.mean\endcsname{0.0}
\expandafter\def\csname val@rev.pol.holm.sspro.p2.rcr-vs-generic-ppi-source.n.p-holm\endcsname{0.876}
\expandafter\def\csname cnt@rev.pol.holm.sspro.p2.rcr-vs-generic-ppi-source.n.p-holm\endcsname{1}
\expandafter\def\csname pct@rev.pol.holm.sspro.p2.rcr-vs-generic-ppi-source.n.p-holm\endcsname{87.6}
\expandafter\def\csname ppt@rev.pol.holm.sspro.p2.rcr-vs-generic-ppi-source.n.p-holm\endcsname{$+$87.6}
\expandafter\def\csname val@rev.pol.holm.sspro.p2.rcr-vs-generic-ppi-source.n.p-holm-text\endcsname{0.8758}
\expandafter\def\csname val@rev.pol.holm.sspro.p2.rcr-vs-generic-ppi-source.n.p-raw\endcsname{0.577}
\expandafter\def\csname cnt@rev.pol.holm.sspro.p2.rcr-vs-generic-ppi-source.n.p-raw\endcsname{1}
\expandafter\def\csname pct@rev.pol.holm.sspro.p2.rcr-vs-generic-ppi-source.n.p-raw\endcsname{57.7}
\expandafter\def\csname ppt@rev.pol.holm.sspro.p2.rcr-vs-generic-ppi-source.n.p-raw\endcsname{$+$57.7}
\expandafter\def\csname val@rev.pol.holm.sspro.p2.rcr-vs-generic-ppi-source.n.reject\endcsname{0}
\expandafter\def\csname cnt@rev.pol.holm.sspro.p2.rcr-vs-generic-ppi-source.n.reject\endcsname{0}
\expandafter\def\csname pct@rev.pol.holm.sspro.p2.rcr-vs-generic-ppi-source.n.reject\endcsname{0.0}
\expandafter\def\csname ppt@rev.pol.holm.sspro.p2.rcr-vs-generic-ppi-source.n.reject\endcsname{0.0}
\expandafter\def\csname val@rev.pol.holm.sspro.p2.rcr-vs-replay-ppi-uniform.f20.lb\endcsname{0.006}
\expandafter\def\csname cnt@rev.pol.holm.sspro.p2.rcr-vs-replay-ppi-uniform.f20.lb\endcsname{0}
\expandafter\def\csname pct@rev.pol.holm.sspro.p2.rcr-vs-replay-ppi-uniform.f20.lb\endcsname{0.6}
\expandafter\def\csname ppt@rev.pol.holm.sspro.p2.rcr-vs-replay-ppi-uniform.f20.lb\endcsname{$+$0.6}
\expandafter\def\csname val@rev.pol.holm.sspro.p2.rcr-vs-replay-ppi-uniform.f20.mean\endcsname{0.010}
\expandafter\def\csname cnt@rev.pol.holm.sspro.p2.rcr-vs-replay-ppi-uniform.f20.mean\endcsname{0}
\expandafter\def\csname pct@rev.pol.holm.sspro.p2.rcr-vs-replay-ppi-uniform.f20.mean\endcsname{1.0}
\expandafter\def\csname ppt@rev.pol.holm.sspro.p2.rcr-vs-replay-ppi-uniform.f20.mean\endcsname{$+$1.0}
\expandafter\def\csname val@rev.pol.holm.sspro.p2.rcr-vs-replay-ppi-uniform.f20.p-holm\endcsname{0.002}
\expandafter\def\csname cnt@rev.pol.holm.sspro.p2.rcr-vs-replay-ppi-uniform.f20.p-holm\endcsname{0}
\expandafter\def\csname pct@rev.pol.holm.sspro.p2.rcr-vs-replay-ppi-uniform.f20.p-holm\endcsname{0.2}
\expandafter\def\csname ppt@rev.pol.holm.sspro.p2.rcr-vs-replay-ppi-uniform.f20.p-holm\endcsname{$+$0.2}
\expandafter\def\csname val@rev.pol.holm.sspro.p2.rcr-vs-replay-ppi-uniform.f20.p-holm-text\endcsname{0.0020}
\expandafter\def\csname val@rev.pol.holm.sspro.p2.rcr-vs-replay-ppi-uniform.f20.p-raw\endcsname{0.000}
\expandafter\def\csname cnt@rev.pol.holm.sspro.p2.rcr-vs-replay-ppi-uniform.f20.p-raw\endcsname{0}
\expandafter\def\csname pct@rev.pol.holm.sspro.p2.rcr-vs-replay-ppi-uniform.f20.p-raw\endcsname{0.0}
\expandafter\def\csname ppt@rev.pol.holm.sspro.p2.rcr-vs-replay-ppi-uniform.f20.p-raw\endcsname{0.0}
\expandafter\def\csname val@rev.pol.holm.sspro.p2.rcr-vs-replay-ppi-uniform.f20.reject\endcsname{1}
\expandafter\def\csname cnt@rev.pol.holm.sspro.p2.rcr-vs-replay-ppi-uniform.f20.reject\endcsname{1}
\expandafter\def\csname pct@rev.pol.holm.sspro.p2.rcr-vs-replay-ppi-uniform.f20.reject\endcsname{100.0}
\expandafter\def\csname ppt@rev.pol.holm.sspro.p2.rcr-vs-replay-ppi-uniform.f20.reject\endcsname{$+$100.0}
\expandafter\def\csname val@rev.pol.holm.sspro.p2.rcr-vs-replay-ppi-uniform.n.lb\endcsname{-0.000}
\expandafter\def\csname cnt@rev.pol.holm.sspro.p2.rcr-vs-replay-ppi-uniform.n.lb\endcsname{0}
\expandafter\def\csname pct@rev.pol.holm.sspro.p2.rcr-vs-replay-ppi-uniform.n.lb\endcsname{0.0}
\expandafter\def\csname ppt@rev.pol.holm.sspro.p2.rcr-vs-replay-ppi-uniform.n.lb\endcsname{0.0}
\expandafter\def\csname val@rev.pol.holm.sspro.p2.rcr-vs-replay-ppi-uniform.n.mean\endcsname{0.000}
\expandafter\def\csname cnt@rev.pol.holm.sspro.p2.rcr-vs-replay-ppi-uniform.n.mean\endcsname{0}
\expandafter\def\csname pct@rev.pol.holm.sspro.p2.rcr-vs-replay-ppi-uniform.n.mean\endcsname{0.0}
\expandafter\def\csname ppt@rev.pol.holm.sspro.p2.rcr-vs-replay-ppi-uniform.n.mean\endcsname{0.0}
\expandafter\def\csname val@rev.pol.holm.sspro.p2.rcr-vs-replay-ppi-uniform.n.p-holm\endcsname{0.876}
\expandafter\def\csname cnt@rev.pol.holm.sspro.p2.rcr-vs-replay-ppi-uniform.n.p-holm\endcsname{1}
\expandafter\def\csname pct@rev.pol.holm.sspro.p2.rcr-vs-replay-ppi-uniform.n.p-holm\endcsname{87.6}
\expandafter\def\csname ppt@rev.pol.holm.sspro.p2.rcr-vs-replay-ppi-uniform.n.p-holm\endcsname{$+$87.6}
\expandafter\def\csname val@rev.pol.holm.sspro.p2.rcr-vs-replay-ppi-uniform.n.p-holm-text\endcsname{0.8758}
\expandafter\def\csname val@rev.pol.holm.sspro.p2.rcr-vs-replay-ppi-uniform.n.p-raw\endcsname{0.438}
\expandafter\def\csname cnt@rev.pol.holm.sspro.p2.rcr-vs-replay-ppi-uniform.n.p-raw\endcsname{0}
\expandafter\def\csname pct@rev.pol.holm.sspro.p2.rcr-vs-replay-ppi-uniform.n.p-raw\endcsname{43.8}
\expandafter\def\csname ppt@rev.pol.holm.sspro.p2.rcr-vs-replay-ppi-uniform.n.p-raw\endcsname{$+$43.8}
\expandafter\def\csname val@rev.pol.holm.sspro.p2.rcr-vs-replay-ppi-uniform.n.reject\endcsname{0}
\expandafter\def\csname cnt@rev.pol.holm.sspro.p2.rcr-vs-replay-ppi-uniform.n.reject\endcsname{0}
\expandafter\def\csname pct@rev.pol.holm.sspro.p2.rcr-vs-replay-ppi-uniform.n.reject\endcsname{0.0}
\expandafter\def\csname ppt@rev.pol.holm.sspro.p2.rcr-vs-replay-ppi-uniform.n.reject\endcsname{0.0}
\expandafter\def\csname val@rev.pol.holm.uiv.p1.rcr-vs-generic-ppi-source.f20.lb\endcsname{0.008}
\expandafter\def\csname cnt@rev.pol.holm.uiv.p1.rcr-vs-generic-ppi-source.f20.lb\endcsname{0}
\expandafter\def\csname pct@rev.pol.holm.uiv.p1.rcr-vs-generic-ppi-source.f20.lb\endcsname{0.8}
\expandafter\def\csname ppt@rev.pol.holm.uiv.p1.rcr-vs-generic-ppi-source.f20.lb\endcsname{$+$0.8}
\expandafter\def\csname val@rev.pol.holm.uiv.p1.rcr-vs-generic-ppi-source.f20.mean\endcsname{0.013}
\expandafter\def\csname cnt@rev.pol.holm.uiv.p1.rcr-vs-generic-ppi-source.f20.mean\endcsname{0}
\expandafter\def\csname pct@rev.pol.holm.uiv.p1.rcr-vs-generic-ppi-source.f20.mean\endcsname{1.3}
\expandafter\def\csname ppt@rev.pol.holm.uiv.p1.rcr-vs-generic-ppi-source.f20.mean\endcsname{$+$1.3}
\expandafter\def\csname val@rev.pol.holm.uiv.p1.rcr-vs-generic-ppi-source.f20.p-holm\endcsname{0.002}
\expandafter\def\csname cnt@rev.pol.holm.uiv.p1.rcr-vs-generic-ppi-source.f20.p-holm\endcsname{0}
\expandafter\def\csname pct@rev.pol.holm.uiv.p1.rcr-vs-generic-ppi-source.f20.p-holm\endcsname{0.2}
\expandafter\def\csname ppt@rev.pol.holm.uiv.p1.rcr-vs-generic-ppi-source.f20.p-holm\endcsname{$+$0.2}
\expandafter\def\csname val@rev.pol.holm.uiv.p1.rcr-vs-generic-ppi-source.f20.p-holm-text\endcsname{0.0020}
\expandafter\def\csname val@rev.pol.holm.uiv.p1.rcr-vs-generic-ppi-source.f20.p-raw\endcsname{0.000}
\expandafter\def\csname cnt@rev.pol.holm.uiv.p1.rcr-vs-generic-ppi-source.f20.p-raw\endcsname{0}
\expandafter\def\csname pct@rev.pol.holm.uiv.p1.rcr-vs-generic-ppi-source.f20.p-raw\endcsname{0.0}
\expandafter\def\csname ppt@rev.pol.holm.uiv.p1.rcr-vs-generic-ppi-source.f20.p-raw\endcsname{0.0}
\expandafter\def\csname val@rev.pol.holm.uiv.p1.rcr-vs-generic-ppi-source.f20.reject\endcsname{1}
\expandafter\def\csname cnt@rev.pol.holm.uiv.p1.rcr-vs-generic-ppi-source.f20.reject\endcsname{1}
\expandafter\def\csname pct@rev.pol.holm.uiv.p1.rcr-vs-generic-ppi-source.f20.reject\endcsname{100.0}
\expandafter\def\csname ppt@rev.pol.holm.uiv.p1.rcr-vs-generic-ppi-source.f20.reject\endcsname{$+$100.0}
\expandafter\def\csname val@rev.pol.holm.uiv.p1.rcr-vs-generic-ppi-source.n.lb\endcsname{0.004}
\expandafter\def\csname cnt@rev.pol.holm.uiv.p1.rcr-vs-generic-ppi-source.n.lb\endcsname{0}
\expandafter\def\csname pct@rev.pol.holm.uiv.p1.rcr-vs-generic-ppi-source.n.lb\endcsname{0.4}
\expandafter\def\csname ppt@rev.pol.holm.uiv.p1.rcr-vs-generic-ppi-source.n.lb\endcsname{$+$0.4}
\expandafter\def\csname val@rev.pol.holm.uiv.p1.rcr-vs-generic-ppi-source.n.mean\endcsname{0.008}
\expandafter\def\csname cnt@rev.pol.holm.uiv.p1.rcr-vs-generic-ppi-source.n.mean\endcsname{0}
\expandafter\def\csname pct@rev.pol.holm.uiv.p1.rcr-vs-generic-ppi-source.n.mean\endcsname{0.8}
\expandafter\def\csname ppt@rev.pol.holm.uiv.p1.rcr-vs-generic-ppi-source.n.mean\endcsname{$+$0.8}
\expandafter\def\csname val@rev.pol.holm.uiv.p1.rcr-vs-generic-ppi-source.n.p-holm\endcsname{0.002}
\expandafter\def\csname cnt@rev.pol.holm.uiv.p1.rcr-vs-generic-ppi-source.n.p-holm\endcsname{0}
\expandafter\def\csname pct@rev.pol.holm.uiv.p1.rcr-vs-generic-ppi-source.n.p-holm\endcsname{0.2}
\expandafter\def\csname ppt@rev.pol.holm.uiv.p1.rcr-vs-generic-ppi-source.n.p-holm\endcsname{$+$0.2}
\expandafter\def\csname val@rev.pol.holm.uiv.p1.rcr-vs-generic-ppi-source.n.p-holm-text\endcsname{0.0020}
\expandafter\def\csname val@rev.pol.holm.uiv.p1.rcr-vs-generic-ppi-source.n.p-raw\endcsname{0.000}
\expandafter\def\csname cnt@rev.pol.holm.uiv.p1.rcr-vs-generic-ppi-source.n.p-raw\endcsname{0}
\expandafter\def\csname pct@rev.pol.holm.uiv.p1.rcr-vs-generic-ppi-source.n.p-raw\endcsname{0.0}
\expandafter\def\csname ppt@rev.pol.holm.uiv.p1.rcr-vs-generic-ppi-source.n.p-raw\endcsname{0.0}
\expandafter\def\csname val@rev.pol.holm.uiv.p1.rcr-vs-generic-ppi-source.n.reject\endcsname{1}
\expandafter\def\csname cnt@rev.pol.holm.uiv.p1.rcr-vs-generic-ppi-source.n.reject\endcsname{1}
\expandafter\def\csname pct@rev.pol.holm.uiv.p1.rcr-vs-generic-ppi-source.n.reject\endcsname{100.0}
\expandafter\def\csname ppt@rev.pol.holm.uiv.p1.rcr-vs-generic-ppi-source.n.reject\endcsname{$+$100.0}
\expandafter\def\csname val@rev.pol.holm.uiv.p1.rcr-vs-replay-ppi-uniform.f20.lb\endcsname{-0.001}
\expandafter\def\csname cnt@rev.pol.holm.uiv.p1.rcr-vs-replay-ppi-uniform.f20.lb\endcsname{0}
\expandafter\def\csname pct@rev.pol.holm.uiv.p1.rcr-vs-replay-ppi-uniform.f20.lb\endcsname{$-$0.1}
\expandafter\def\csname ppt@rev.pol.holm.uiv.p1.rcr-vs-replay-ppi-uniform.f20.lb\endcsname{$-$0.1}
\expandafter\def\csname val@rev.pol.holm.uiv.p1.rcr-vs-replay-ppi-uniform.f20.mean\endcsname{0.001}
\expandafter\def\csname cnt@rev.pol.holm.uiv.p1.rcr-vs-replay-ppi-uniform.f20.mean\endcsname{0}
\expandafter\def\csname pct@rev.pol.holm.uiv.p1.rcr-vs-replay-ppi-uniform.f20.mean\endcsname{0.1}
\expandafter\def\csname ppt@rev.pol.holm.uiv.p1.rcr-vs-replay-ppi-uniform.f20.mean\endcsname{$+$0.1}
\expandafter\def\csname val@rev.pol.holm.uiv.p1.rcr-vs-replay-ppi-uniform.f20.p-holm\endcsname{0.165}
\expandafter\def\csname cnt@rev.pol.holm.uiv.p1.rcr-vs-replay-ppi-uniform.f20.p-holm\endcsname{0}
\expandafter\def\csname pct@rev.pol.holm.uiv.p1.rcr-vs-replay-ppi-uniform.f20.p-holm\endcsname{16.5}
\expandafter\def\csname ppt@rev.pol.holm.uiv.p1.rcr-vs-replay-ppi-uniform.f20.p-holm\endcsname{$+$16.5}
\expandafter\def\csname val@rev.pol.holm.uiv.p1.rcr-vs-replay-ppi-uniform.f20.p-holm-text\endcsname{0.1647}
\expandafter\def\csname val@rev.pol.holm.uiv.p1.rcr-vs-replay-ppi-uniform.f20.p-raw\endcsname{0.165}
\expandafter\def\csname cnt@rev.pol.holm.uiv.p1.rcr-vs-replay-ppi-uniform.f20.p-raw\endcsname{0}
\expandafter\def\csname pct@rev.pol.holm.uiv.p1.rcr-vs-replay-ppi-uniform.f20.p-raw\endcsname{16.5}
\expandafter\def\csname ppt@rev.pol.holm.uiv.p1.rcr-vs-replay-ppi-uniform.f20.p-raw\endcsname{$+$16.5}
\expandafter\def\csname val@rev.pol.holm.uiv.p1.rcr-vs-replay-ppi-uniform.f20.reject\endcsname{0}
\expandafter\def\csname cnt@rev.pol.holm.uiv.p1.rcr-vs-replay-ppi-uniform.f20.reject\endcsname{0}
\expandafter\def\csname pct@rev.pol.holm.uiv.p1.rcr-vs-replay-ppi-uniform.f20.reject\endcsname{0.0}
\expandafter\def\csname ppt@rev.pol.holm.uiv.p1.rcr-vs-replay-ppi-uniform.f20.reject\endcsname{0.0}
\expandafter\def\csname val@rev.pol.holm.uiv.p1.rcr-vs-replay-ppi-uniform.n.lb\endcsname{0.002}
\expandafter\def\csname cnt@rev.pol.holm.uiv.p1.rcr-vs-replay-ppi-uniform.n.lb\endcsname{0}
\expandafter\def\csname pct@rev.pol.holm.uiv.p1.rcr-vs-replay-ppi-uniform.n.lb\endcsname{0.2}
\expandafter\def\csname ppt@rev.pol.holm.uiv.p1.rcr-vs-replay-ppi-uniform.n.lb\endcsname{$+$0.2}
\expandafter\def\csname val@rev.pol.holm.uiv.p1.rcr-vs-replay-ppi-uniform.n.mean\endcsname{0.004}
\expandafter\def\csname cnt@rev.pol.holm.uiv.p1.rcr-vs-replay-ppi-uniform.n.mean\endcsname{0}
\expandafter\def\csname pct@rev.pol.holm.uiv.p1.rcr-vs-replay-ppi-uniform.n.mean\endcsname{0.4}
\expandafter\def\csname ppt@rev.pol.holm.uiv.p1.rcr-vs-replay-ppi-uniform.n.mean\endcsname{$+$0.4}
\expandafter\def\csname val@rev.pol.holm.uiv.p1.rcr-vs-replay-ppi-uniform.n.p-holm\endcsname{0.002}
\expandafter\def\csname cnt@rev.pol.holm.uiv.p1.rcr-vs-replay-ppi-uniform.n.p-holm\endcsname{0}
\expandafter\def\csname pct@rev.pol.holm.uiv.p1.rcr-vs-replay-ppi-uniform.n.p-holm\endcsname{0.2}
\expandafter\def\csname ppt@rev.pol.holm.uiv.p1.rcr-vs-replay-ppi-uniform.n.p-holm\endcsname{$+$0.2}
\expandafter\def\csname val@rev.pol.holm.uiv.p1.rcr-vs-replay-ppi-uniform.n.p-holm-text\endcsname{0.0020}
\expandafter\def\csname val@rev.pol.holm.uiv.p1.rcr-vs-replay-ppi-uniform.n.p-raw\endcsname{0.000}
\expandafter\def\csname cnt@rev.pol.holm.uiv.p1.rcr-vs-replay-ppi-uniform.n.p-raw\endcsname{0}
\expandafter\def\csname pct@rev.pol.holm.uiv.p1.rcr-vs-replay-ppi-uniform.n.p-raw\endcsname{0.0}
\expandafter\def\csname ppt@rev.pol.holm.uiv.p1.rcr-vs-replay-ppi-uniform.n.p-raw\endcsname{0.0}
\expandafter\def\csname val@rev.pol.holm.uiv.p1.rcr-vs-replay-ppi-uniform.n.reject\endcsname{1}
\expandafter\def\csname cnt@rev.pol.holm.uiv.p1.rcr-vs-replay-ppi-uniform.n.reject\endcsname{1}
\expandafter\def\csname pct@rev.pol.holm.uiv.p1.rcr-vs-replay-ppi-uniform.n.reject\endcsname{100.0}
\expandafter\def\csname ppt@rev.pol.holm.uiv.p1.rcr-vs-replay-ppi-uniform.n.reject\endcsname{$+$100.0}
\expandafter\def\csname val@rev.pol.holm.uiv.p2.rcr-vs-generic-ppi-source.f20.lb\endcsname{0.001}
\expandafter\def\csname cnt@rev.pol.holm.uiv.p2.rcr-vs-generic-ppi-source.f20.lb\endcsname{0}
\expandafter\def\csname pct@rev.pol.holm.uiv.p2.rcr-vs-generic-ppi-source.f20.lb\endcsname{0.1}
\expandafter\def\csname ppt@rev.pol.holm.uiv.p2.rcr-vs-generic-ppi-source.f20.lb\endcsname{$+$0.1}
\expandafter\def\csname val@rev.pol.holm.uiv.p2.rcr-vs-generic-ppi-source.f20.mean\endcsname{0.004}
\expandafter\def\csname cnt@rev.pol.holm.uiv.p2.rcr-vs-generic-ppi-source.f20.mean\endcsname{0}
\expandafter\def\csname pct@rev.pol.holm.uiv.p2.rcr-vs-generic-ppi-source.f20.mean\endcsname{0.4}
\expandafter\def\csname ppt@rev.pol.holm.uiv.p2.rcr-vs-generic-ppi-source.f20.mean\endcsname{$+$0.4}
\expandafter\def\csname val@rev.pol.holm.uiv.p2.rcr-vs-generic-ppi-source.f20.p-holm\endcsname{0.049}
\expandafter\def\csname cnt@rev.pol.holm.uiv.p2.rcr-vs-generic-ppi-source.f20.p-holm\endcsname{0}
\expandafter\def\csname pct@rev.pol.holm.uiv.p2.rcr-vs-generic-ppi-source.f20.p-holm\endcsname{4.9}
\expandafter\def\csname ppt@rev.pol.holm.uiv.p2.rcr-vs-generic-ppi-source.f20.p-holm\endcsname{$+$4.9}
\expandafter\def\csname val@rev.pol.holm.uiv.p2.rcr-vs-generic-ppi-source.f20.p-holm-text\endcsname{0.0490}
\expandafter\def\csname val@rev.pol.holm.uiv.p2.rcr-vs-generic-ppi-source.f20.p-raw\endcsname{0.012}
\expandafter\def\csname cnt@rev.pol.holm.uiv.p2.rcr-vs-generic-ppi-source.f20.p-raw\endcsname{0}
\expandafter\def\csname pct@rev.pol.holm.uiv.p2.rcr-vs-generic-ppi-source.f20.p-raw\endcsname{1.2}
\expandafter\def\csname ppt@rev.pol.holm.uiv.p2.rcr-vs-generic-ppi-source.f20.p-raw\endcsname{$+$1.2}
\expandafter\def\csname val@rev.pol.holm.uiv.p2.rcr-vs-generic-ppi-source.f20.reject\endcsname{1}
\expandafter\def\csname cnt@rev.pol.holm.uiv.p2.rcr-vs-generic-ppi-source.f20.reject\endcsname{1}
\expandafter\def\csname pct@rev.pol.holm.uiv.p2.rcr-vs-generic-ppi-source.f20.reject\endcsname{100.0}
\expandafter\def\csname ppt@rev.pol.holm.uiv.p2.rcr-vs-generic-ppi-source.f20.reject\endcsname{$+$100.0}
\expandafter\def\csname val@rev.pol.holm.uiv.p2.rcr-vs-generic-ppi-source.n.lb\endcsname{0.000}
\expandafter\def\csname cnt@rev.pol.holm.uiv.p2.rcr-vs-generic-ppi-source.n.lb\endcsname{0}
\expandafter\def\csname pct@rev.pol.holm.uiv.p2.rcr-vs-generic-ppi-source.n.lb\endcsname{0.0}
\expandafter\def\csname ppt@rev.pol.holm.uiv.p2.rcr-vs-generic-ppi-source.n.lb\endcsname{0.0}
\expandafter\def\csname val@rev.pol.holm.uiv.p2.rcr-vs-generic-ppi-source.n.mean\endcsname{0.001}
\expandafter\def\csname cnt@rev.pol.holm.uiv.p2.rcr-vs-generic-ppi-source.n.mean\endcsname{0}
\expandafter\def\csname pct@rev.pol.holm.uiv.p2.rcr-vs-generic-ppi-source.n.mean\endcsname{0.1}
\expandafter\def\csname ppt@rev.pol.holm.uiv.p2.rcr-vs-generic-ppi-source.n.mean\endcsname{$+$0.1}
\expandafter\def\csname val@rev.pol.holm.uiv.p2.rcr-vs-generic-ppi-source.n.p-holm\endcsname{0.137}
\expandafter\def\csname cnt@rev.pol.holm.uiv.p2.rcr-vs-generic-ppi-source.n.p-holm\endcsname{0}
\expandafter\def\csname pct@rev.pol.holm.uiv.p2.rcr-vs-generic-ppi-source.n.p-holm\endcsname{13.7}
\expandafter\def\csname ppt@rev.pol.holm.uiv.p2.rcr-vs-generic-ppi-source.n.p-holm\endcsname{$+$13.7}
\expandafter\def\csname val@rev.pol.holm.uiv.p2.rcr-vs-generic-ppi-source.n.p-holm-text\endcsname{0.1372}
\expandafter\def\csname val@rev.pol.holm.uiv.p2.rcr-vs-generic-ppi-source.n.p-raw\endcsname{0.046}
\expandafter\def\csname cnt@rev.pol.holm.uiv.p2.rcr-vs-generic-ppi-source.n.p-raw\endcsname{0}
\expandafter\def\csname pct@rev.pol.holm.uiv.p2.rcr-vs-generic-ppi-source.n.p-raw\endcsname{4.6}
\expandafter\def\csname ppt@rev.pol.holm.uiv.p2.rcr-vs-generic-ppi-source.n.p-raw\endcsname{$+$4.6}
\expandafter\def\csname val@rev.pol.holm.uiv.p2.rcr-vs-generic-ppi-source.n.reject\endcsname{0}
\expandafter\def\csname cnt@rev.pol.holm.uiv.p2.rcr-vs-generic-ppi-source.n.reject\endcsname{0}
\expandafter\def\csname pct@rev.pol.holm.uiv.p2.rcr-vs-generic-ppi-source.n.reject\endcsname{0.0}
\expandafter\def\csname ppt@rev.pol.holm.uiv.p2.rcr-vs-generic-ppi-source.n.reject\endcsname{0.0}
\expandafter\def\csname val@rev.pol.holm.uiv.p2.rcr-vs-replay-ppi-uniform.f20.lb\endcsname{0.000}
\expandafter\def\csname cnt@rev.pol.holm.uiv.p2.rcr-vs-replay-ppi-uniform.f20.lb\endcsname{0}
\expandafter\def\csname pct@rev.pol.holm.uiv.p2.rcr-vs-replay-ppi-uniform.f20.lb\endcsname{0.0}
\expandafter\def\csname ppt@rev.pol.holm.uiv.p2.rcr-vs-replay-ppi-uniform.f20.lb\endcsname{0.0}
\expandafter\def\csname val@rev.pol.holm.uiv.p2.rcr-vs-replay-ppi-uniform.f20.mean\endcsname{0.002}
\expandafter\def\csname cnt@rev.pol.holm.uiv.p2.rcr-vs-replay-ppi-uniform.f20.mean\endcsname{0}
\expandafter\def\csname pct@rev.pol.holm.uiv.p2.rcr-vs-replay-ppi-uniform.f20.mean\endcsname{0.2}
\expandafter\def\csname ppt@rev.pol.holm.uiv.p2.rcr-vs-replay-ppi-uniform.f20.mean\endcsname{$+$0.2}
\expandafter\def\csname val@rev.pol.holm.uiv.p2.rcr-vs-replay-ppi-uniform.f20.p-holm\endcsname{0.049}
\expandafter\def\csname cnt@rev.pol.holm.uiv.p2.rcr-vs-replay-ppi-uniform.f20.p-holm\endcsname{0}
\expandafter\def\csname pct@rev.pol.holm.uiv.p2.rcr-vs-replay-ppi-uniform.f20.p-holm\endcsname{4.9}
\expandafter\def\csname ppt@rev.pol.holm.uiv.p2.rcr-vs-replay-ppi-uniform.f20.p-holm\endcsname{$+$4.9}
\expandafter\def\csname val@rev.pol.holm.uiv.p2.rcr-vs-replay-ppi-uniform.f20.p-holm-text\endcsname{0.0490}
\expandafter\def\csname val@rev.pol.holm.uiv.p2.rcr-vs-replay-ppi-uniform.f20.p-raw\endcsname{0.013}
\expandafter\def\csname cnt@rev.pol.holm.uiv.p2.rcr-vs-replay-ppi-uniform.f20.p-raw\endcsname{0}
\expandafter\def\csname pct@rev.pol.holm.uiv.p2.rcr-vs-replay-ppi-uniform.f20.p-raw\endcsname{1.3}
\expandafter\def\csname ppt@rev.pol.holm.uiv.p2.rcr-vs-replay-ppi-uniform.f20.p-raw\endcsname{$+$1.3}
\expandafter\def\csname val@rev.pol.holm.uiv.p2.rcr-vs-replay-ppi-uniform.f20.reject\endcsname{1}
\expandafter\def\csname cnt@rev.pol.holm.uiv.p2.rcr-vs-replay-ppi-uniform.f20.reject\endcsname{1}
\expandafter\def\csname pct@rev.pol.holm.uiv.p2.rcr-vs-replay-ppi-uniform.f20.reject\endcsname{100.0}
\expandafter\def\csname ppt@rev.pol.holm.uiv.p2.rcr-vs-replay-ppi-uniform.f20.reject\endcsname{$+$100.0}
\expandafter\def\csname val@rev.pol.holm.uiv.p2.rcr-vs-replay-ppi-uniform.n.lb\endcsname{0.000}
\expandafter\def\csname cnt@rev.pol.holm.uiv.p2.rcr-vs-replay-ppi-uniform.n.lb\endcsname{0}
\expandafter\def\csname pct@rev.pol.holm.uiv.p2.rcr-vs-replay-ppi-uniform.n.lb\endcsname{0.0}
\expandafter\def\csname ppt@rev.pol.holm.uiv.p2.rcr-vs-replay-ppi-uniform.n.lb\endcsname{0.0}
\expandafter\def\csname val@rev.pol.holm.uiv.p2.rcr-vs-replay-ppi-uniform.n.mean\endcsname{0.001}
\expandafter\def\csname cnt@rev.pol.holm.uiv.p2.rcr-vs-replay-ppi-uniform.n.mean\endcsname{0}
\expandafter\def\csname pct@rev.pol.holm.uiv.p2.rcr-vs-replay-ppi-uniform.n.mean\endcsname{0.1}
\expandafter\def\csname ppt@rev.pol.holm.uiv.p2.rcr-vs-replay-ppi-uniform.n.mean\endcsname{$+$0.1}
\expandafter\def\csname val@rev.pol.holm.uiv.p2.rcr-vs-replay-ppi-uniform.n.p-holm\endcsname{0.021}
\expandafter\def\csname cnt@rev.pol.holm.uiv.p2.rcr-vs-replay-ppi-uniform.n.p-holm\endcsname{0}
\expandafter\def\csname pct@rev.pol.holm.uiv.p2.rcr-vs-replay-ppi-uniform.n.p-holm\endcsname{2.1}
\expandafter\def\csname ppt@rev.pol.holm.uiv.p2.rcr-vs-replay-ppi-uniform.n.p-holm\endcsname{$+$2.1}
\expandafter\def\csname val@rev.pol.holm.uiv.p2.rcr-vs-replay-ppi-uniform.n.p-holm-text\endcsname{0.0210}
\expandafter\def\csname val@rev.pol.holm.uiv.p2.rcr-vs-replay-ppi-uniform.n.p-raw\endcsname{0.003}
\expandafter\def\csname cnt@rev.pol.holm.uiv.p2.rcr-vs-replay-ppi-uniform.n.p-raw\endcsname{0}
\expandafter\def\csname pct@rev.pol.holm.uiv.p2.rcr-vs-replay-ppi-uniform.n.p-raw\endcsname{0.3}
\expandafter\def\csname ppt@rev.pol.holm.uiv.p2.rcr-vs-replay-ppi-uniform.n.p-raw\endcsname{$+$0.3}
\expandafter\def\csname val@rev.pol.holm.uiv.p2.rcr-vs-replay-ppi-uniform.n.reject\endcsname{1}
\expandafter\def\csname cnt@rev.pol.holm.uiv.p2.rcr-vs-replay-ppi-uniform.n.reject\endcsname{1}
\expandafter\def\csname pct@rev.pol.holm.uiv.p2.rcr-vs-replay-ppi-uniform.n.reject\endcsname{100.0}
\expandafter\def\csname ppt@rev.pol.holm.uiv.p2.rcr-vs-replay-ppi-uniform.n.reject\endcsname{$+$100.0}
\expandafter\def\csname val@rev.pol.infeas.claim.excl20\endcsname{4}
\expandafter\def\csname cnt@rev.pol.infeas.claim.excl20\endcsname{4}
\expandafter\def\csname pct@rev.pol.infeas.claim.excl20\endcsname{400.0}
\expandafter\def\csname ppt@rev.pol.infeas.claim.excl20\endcsname{$+$400.0}
\expandafter\def\csname val@rev.pol.infeas.claim.exclN\endcsname{4}
\expandafter\def\csname cnt@rev.pol.infeas.claim.exclN\endcsname{4}
\expandafter\def\csname pct@rev.pol.infeas.claim.exclN\endcsname{400.0}
\expandafter\def\csname ppt@rev.pol.infeas.claim.exclN\endcsname{$+$400.0}
\expandafter\def\csname val@rev.pol.infeas.claim.incl20\endcsname{4}
\expandafter\def\csname cnt@rev.pol.infeas.claim.incl20\endcsname{4}
\expandafter\def\csname pct@rev.pol.infeas.claim.incl20\endcsname{400.0}
\expandafter\def\csname ppt@rev.pol.infeas.claim.incl20\endcsname{$+$400.0}
\expandafter\def\csname val@rev.pol.infeas.claim.inclN\endcsname{4}
\expandafter\def\csname cnt@rev.pol.infeas.claim.inclN\endcsname{4}
\expandafter\def\csname pct@rev.pol.infeas.claim.inclN\endcsname{400.0}
\expandafter\def\csname ppt@rev.pol.infeas.claim.inclN\endcsname{$+$400.0}
\expandafter\def\csname val@rev.pol.infeas.claim.sspro.p1.margin20\endcsname{0.000}
\expandafter\def\csname cnt@rev.pol.infeas.claim.sspro.p1.margin20\endcsname{0}
\expandafter\def\csname pct@rev.pol.infeas.claim.sspro.p1.margin20\endcsname{0.0}
\expandafter\def\csname ppt@rev.pol.infeas.claim.sspro.p1.margin20\endcsname{0.0}
\expandafter\def\csname val@rev.pol.infeas.claim.sspro.p1.marginN\endcsname{0.000}
\expandafter\def\csname cnt@rev.pol.infeas.claim.sspro.p1.marginN\endcsname{0}
\expandafter\def\csname pct@rev.pol.infeas.claim.sspro.p1.marginN\endcsname{0.0}
\expandafter\def\csname ppt@rev.pol.infeas.claim.sspro.p1.marginN\endcsname{0.0}
\expandafter\def\csname val@rev.pol.infeas.claim.sspro.p2.margin20\endcsname{0.000}
\expandafter\def\csname cnt@rev.pol.infeas.claim.sspro.p2.margin20\endcsname{0}
\expandafter\def\csname pct@rev.pol.infeas.claim.sspro.p2.margin20\endcsname{0.0}
\expandafter\def\csname ppt@rev.pol.infeas.claim.sspro.p2.margin20\endcsname{0.0}
\expandafter\def\csname val@rev.pol.infeas.claim.sspro.p2.marginN\endcsname{0.000}
\expandafter\def\csname cnt@rev.pol.infeas.claim.sspro.p2.marginN\endcsname{0}
\expandafter\def\csname pct@rev.pol.infeas.claim.sspro.p2.marginN\endcsname{0.0}
\expandafter\def\csname ppt@rev.pol.infeas.claim.sspro.p2.marginN\endcsname{0.0}
\expandafter\def\csname val@rev.pol.infeas.claim.uiv.p1.margin20\endcsname{0.023}
\expandafter\def\csname cnt@rev.pol.infeas.claim.uiv.p1.margin20\endcsname{0}
\expandafter\def\csname pct@rev.pol.infeas.claim.uiv.p1.margin20\endcsname{2.3}
\expandafter\def\csname ppt@rev.pol.infeas.claim.uiv.p1.margin20\endcsname{$+$2.3}
\expandafter\def\csname val@rev.pol.infeas.claim.uiv.p1.marginN\endcsname{0.028}
\expandafter\def\csname cnt@rev.pol.infeas.claim.uiv.p1.marginN\endcsname{0}
\expandafter\def\csname pct@rev.pol.infeas.claim.uiv.p1.marginN\endcsname{2.8}
\expandafter\def\csname ppt@rev.pol.infeas.claim.uiv.p1.marginN\endcsname{$+$2.8}
\expandafter\def\csname val@rev.pol.infeas.claim.uiv.p2.margin20\endcsname{0.001}
\expandafter\def\csname cnt@rev.pol.infeas.claim.uiv.p2.margin20\endcsname{0}
\expandafter\def\csname pct@rev.pol.infeas.claim.uiv.p2.margin20\endcsname{0.1}
\expandafter\def\csname ppt@rev.pol.infeas.claim.uiv.p2.margin20\endcsname{$+$0.1}
\expandafter\def\csname val@rev.pol.infeas.claim.uiv.p2.marginN\endcsname{0.002}
\expandafter\def\csname cnt@rev.pol.infeas.claim.uiv.p2.marginN\endcsname{0}
\expandafter\def\csname pct@rev.pol.infeas.claim.uiv.p2.marginN\endcsname{0.2}
\expandafter\def\csname ppt@rev.pol.infeas.claim.uiv.p2.marginN\endcsname{$+$0.2}
\expandafter\def\csname val@rev.pol.infeas.n-settings\endcsname{4}
\expandafter\def\csname cnt@rev.pol.infeas.n-settings\endcsname{4}
\expandafter\def\csname pct@rev.pol.infeas.n-settings\endcsname{400.0}
\expandafter\def\csname ppt@rev.pol.infeas.n-settings\endcsname{$+$400.0}
\expandafter\def\csname val@rev.pol.infeas.sspro.p1@0.1.n-skills\endcsname{4}
\expandafter\def\csname cnt@rev.pol.infeas.sspro.p1@0.1.n-skills\endcsname{4}
\expandafter\def\csname pct@rev.pol.infeas.sspro.p1@0.1.n-skills\endcsname{400.0}
\expandafter\def\csname ppt@rev.pol.infeas.sspro.p1@0.1.n-skills\endcsname{$+$400.0}
\expandafter\def\csname val@rev.pol.infeas.sspro.p1@0.1.n20\endcsname{20}
\expandafter\def\csname cnt@rev.pol.infeas.sspro.p1@0.1.n20\endcsname{20}
\expandafter\def\csname pct@rev.pol.infeas.sspro.p1@0.1.n20\endcsname{2000.0}
\expandafter\def\csname ppt@rev.pol.infeas.sspro.p1@0.1.n20\endcsname{$+$2000.0}
\expandafter\def\csname val@rev.pol.infeas.sspro.p1@0.1.nN\endcsname{10000}
\expandafter\def\csname cnt@rev.pol.infeas.sspro.p1@0.1.nN\endcsname{10{,}000}
\expandafter\def\csname pct@rev.pol.infeas.sspro.p1@0.1.nN\endcsname{1000000.0}
\expandafter\def\csname ppt@rev.pol.infeas.sspro.p1@0.1.nN\endcsname{$+$1000000.0}
\expandafter\def\csname val@rev.pol.infeas.sspro.p1@0.25.n-skills\endcsname{0}
\expandafter\def\csname cnt@rev.pol.infeas.sspro.p1@0.25.n-skills\endcsname{0}
\expandafter\def\csname pct@rev.pol.infeas.sspro.p1@0.25.n-skills\endcsname{0.0}
\expandafter\def\csname ppt@rev.pol.infeas.sspro.p1@0.25.n-skills\endcsname{0.0}
\expandafter\def\csname val@rev.pol.infeas.sspro.p1@0.25.n20\endcsname{0}
\expandafter\def\csname cnt@rev.pol.infeas.sspro.p1@0.25.n20\endcsname{0}
\expandafter\def\csname pct@rev.pol.infeas.sspro.p1@0.25.n20\endcsname{0.0}
\expandafter\def\csname ppt@rev.pol.infeas.sspro.p1@0.25.n20\endcsname{0.0}
\expandafter\def\csname val@rev.pol.infeas.sspro.p1@0.25.nN\endcsname{0}
\expandafter\def\csname cnt@rev.pol.infeas.sspro.p1@0.25.nN\endcsname{0}
\expandafter\def\csname pct@rev.pol.infeas.sspro.p1@0.25.nN\endcsname{0.0}
\expandafter\def\csname ppt@rev.pol.infeas.sspro.p1@0.25.nN\endcsname{0.0}
\expandafter\def\csname val@rev.pol.infeas.sspro.p1@0.5.n-skills\endcsname{0}
\expandafter\def\csname cnt@rev.pol.infeas.sspro.p1@0.5.n-skills\endcsname{0}
\expandafter\def\csname pct@rev.pol.infeas.sspro.p1@0.5.n-skills\endcsname{0.0}
\expandafter\def\csname ppt@rev.pol.infeas.sspro.p1@0.5.n-skills\endcsname{0.0}
\expandafter\def\csname val@rev.pol.infeas.sspro.p1@0.5.n20\endcsname{0}
\expandafter\def\csname cnt@rev.pol.infeas.sspro.p1@0.5.n20\endcsname{0}
\expandafter\def\csname pct@rev.pol.infeas.sspro.p1@0.5.n20\endcsname{0.0}
\expandafter\def\csname ppt@rev.pol.infeas.sspro.p1@0.5.n20\endcsname{0.0}
\expandafter\def\csname val@rev.pol.infeas.sspro.p1@0.5.nN\endcsname{0}
\expandafter\def\csname cnt@rev.pol.infeas.sspro.p1@0.5.nN\endcsname{0}
\expandafter\def\csname pct@rev.pol.infeas.sspro.p1@0.5.nN\endcsname{0.0}
\expandafter\def\csname ppt@rev.pol.infeas.sspro.p1@0.5.nN\endcsname{0.0}
\expandafter\def\csname val@rev.pol.infeas.sspro.p1@1.0.n-skills\endcsname{0}
\expandafter\def\csname cnt@rev.pol.infeas.sspro.p1@1.0.n-skills\endcsname{0}
\expandafter\def\csname pct@rev.pol.infeas.sspro.p1@1.0.n-skills\endcsname{0.0}
\expandafter\def\csname ppt@rev.pol.infeas.sspro.p1@1.0.n-skills\endcsname{0.0}
\expandafter\def\csname val@rev.pol.infeas.sspro.p1@1.0.n20\endcsname{0}
\expandafter\def\csname cnt@rev.pol.infeas.sspro.p1@1.0.n20\endcsname{0}
\expandafter\def\csname pct@rev.pol.infeas.sspro.p1@1.0.n20\endcsname{0.0}
\expandafter\def\csname ppt@rev.pol.infeas.sspro.p1@1.0.n20\endcsname{0.0}
\expandafter\def\csname val@rev.pol.infeas.sspro.p1@1.0.nN\endcsname{0}
\expandafter\def\csname cnt@rev.pol.infeas.sspro.p1@1.0.nN\endcsname{0}
\expandafter\def\csname pct@rev.pol.infeas.sspro.p1@1.0.nN\endcsname{0.0}
\expandafter\def\csname ppt@rev.pol.infeas.sspro.p1@1.0.nN\endcsname{0.0}
\expandafter\def\csname val@rev.pol.infeas.sspro.p2@0.1.n-skills\endcsname{1}
\expandafter\def\csname cnt@rev.pol.infeas.sspro.p2@0.1.n-skills\endcsname{1}
\expandafter\def\csname pct@rev.pol.infeas.sspro.p2@0.1.n-skills\endcsname{100.0}
\expandafter\def\csname ppt@rev.pol.infeas.sspro.p2@0.1.n-skills\endcsname{$+$100.0}
\expandafter\def\csname val@rev.pol.infeas.sspro.p2@0.1.n20\endcsname{20}
\expandafter\def\csname cnt@rev.pol.infeas.sspro.p2@0.1.n20\endcsname{20}
\expandafter\def\csname pct@rev.pol.infeas.sspro.p2@0.1.n20\endcsname{2000.0}
\expandafter\def\csname ppt@rev.pol.infeas.sspro.p2@0.1.n20\endcsname{$+$2000.0}
\expandafter\def\csname val@rev.pol.infeas.sspro.p2@0.1.nN\endcsname{10000}
\expandafter\def\csname cnt@rev.pol.infeas.sspro.p2@0.1.nN\endcsname{10{,}000}
\expandafter\def\csname pct@rev.pol.infeas.sspro.p2@0.1.nN\endcsname{1000000.0}
\expandafter\def\csname ppt@rev.pol.infeas.sspro.p2@0.1.nN\endcsname{$+$1000000.0}
\expandafter\def\csname val@rev.pol.infeas.sspro.p2@0.25.n-skills\endcsname{0}
\expandafter\def\csname cnt@rev.pol.infeas.sspro.p2@0.25.n-skills\endcsname{0}
\expandafter\def\csname pct@rev.pol.infeas.sspro.p2@0.25.n-skills\endcsname{0.0}
\expandafter\def\csname ppt@rev.pol.infeas.sspro.p2@0.25.n-skills\endcsname{0.0}
\expandafter\def\csname val@rev.pol.infeas.sspro.p2@0.25.n20\endcsname{0}
\expandafter\def\csname cnt@rev.pol.infeas.sspro.p2@0.25.n20\endcsname{0}
\expandafter\def\csname pct@rev.pol.infeas.sspro.p2@0.25.n20\endcsname{0.0}
\expandafter\def\csname ppt@rev.pol.infeas.sspro.p2@0.25.n20\endcsname{0.0}
\expandafter\def\csname val@rev.pol.infeas.sspro.p2@0.25.nN\endcsname{0}
\expandafter\def\csname cnt@rev.pol.infeas.sspro.p2@0.25.nN\endcsname{0}
\expandafter\def\csname pct@rev.pol.infeas.sspro.p2@0.25.nN\endcsname{0.0}
\expandafter\def\csname ppt@rev.pol.infeas.sspro.p2@0.25.nN\endcsname{0.0}
\expandafter\def\csname val@rev.pol.infeas.sspro.p2@0.5.n-skills\endcsname{0}
\expandafter\def\csname cnt@rev.pol.infeas.sspro.p2@0.5.n-skills\endcsname{0}
\expandafter\def\csname pct@rev.pol.infeas.sspro.p2@0.5.n-skills\endcsname{0.0}
\expandafter\def\csname ppt@rev.pol.infeas.sspro.p2@0.5.n-skills\endcsname{0.0}
\expandafter\def\csname val@rev.pol.infeas.sspro.p2@0.5.n20\endcsname{0}
\expandafter\def\csname cnt@rev.pol.infeas.sspro.p2@0.5.n20\endcsname{0}
\expandafter\def\csname pct@rev.pol.infeas.sspro.p2@0.5.n20\endcsname{0.0}
\expandafter\def\csname ppt@rev.pol.infeas.sspro.p2@0.5.n20\endcsname{0.0}
\expandafter\def\csname val@rev.pol.infeas.sspro.p2@0.5.nN\endcsname{0}
\expandafter\def\csname cnt@rev.pol.infeas.sspro.p2@0.5.nN\endcsname{0}
\expandafter\def\csname pct@rev.pol.infeas.sspro.p2@0.5.nN\endcsname{0.0}
\expandafter\def\csname ppt@rev.pol.infeas.sspro.p2@0.5.nN\endcsname{0.0}
\expandafter\def\csname val@rev.pol.infeas.sspro.p2@1.0.n-skills\endcsname{0}
\expandafter\def\csname cnt@rev.pol.infeas.sspro.p2@1.0.n-skills\endcsname{0}
\expandafter\def\csname pct@rev.pol.infeas.sspro.p2@1.0.n-skills\endcsname{0.0}
\expandafter\def\csname ppt@rev.pol.infeas.sspro.p2@1.0.n-skills\endcsname{0.0}
\expandafter\def\csname val@rev.pol.infeas.sspro.p2@1.0.n20\endcsname{0}
\expandafter\def\csname cnt@rev.pol.infeas.sspro.p2@1.0.n20\endcsname{0}
\expandafter\def\csname pct@rev.pol.infeas.sspro.p2@1.0.n20\endcsname{0.0}
\expandafter\def\csname ppt@rev.pol.infeas.sspro.p2@1.0.n20\endcsname{0.0}
\expandafter\def\csname val@rev.pol.infeas.sspro.p2@1.0.nN\endcsname{0}
\expandafter\def\csname cnt@rev.pol.infeas.sspro.p2@1.0.nN\endcsname{0}
\expandafter\def\csname pct@rev.pol.infeas.sspro.p2@1.0.nN\endcsname{0.0}
\expandafter\def\csname ppt@rev.pol.infeas.sspro.p2@1.0.nN\endcsname{0.0}
\expandafter\def\csname val@rev.pol.infeas.uiv.p1@0.1.n-skills\endcsname{0}
\expandafter\def\csname cnt@rev.pol.infeas.uiv.p1@0.1.n-skills\endcsname{0}
\expandafter\def\csname pct@rev.pol.infeas.uiv.p1@0.1.n-skills\endcsname{0.0}
\expandafter\def\csname ppt@rev.pol.infeas.uiv.p1@0.1.n-skills\endcsname{0.0}
\expandafter\def\csname val@rev.pol.infeas.uiv.p1@0.1.n20\endcsname{0}
\expandafter\def\csname cnt@rev.pol.infeas.uiv.p1@0.1.n20\endcsname{0}
\expandafter\def\csname pct@rev.pol.infeas.uiv.p1@0.1.n20\endcsname{0.0}
\expandafter\def\csname ppt@rev.pol.infeas.uiv.p1@0.1.n20\endcsname{0.0}
\expandafter\def\csname val@rev.pol.infeas.uiv.p1@0.1.nN\endcsname{0}
\expandafter\def\csname cnt@rev.pol.infeas.uiv.p1@0.1.nN\endcsname{0}
\expandafter\def\csname pct@rev.pol.infeas.uiv.p1@0.1.nN\endcsname{0.0}
\expandafter\def\csname ppt@rev.pol.infeas.uiv.p1@0.1.nN\endcsname{0.0}
\expandafter\def\csname val@rev.pol.infeas.uiv.p1@0.25.n-skills\endcsname{0}
\expandafter\def\csname cnt@rev.pol.infeas.uiv.p1@0.25.n-skills\endcsname{0}
\expandafter\def\csname pct@rev.pol.infeas.uiv.p1@0.25.n-skills\endcsname{0.0}
\expandafter\def\csname ppt@rev.pol.infeas.uiv.p1@0.25.n-skills\endcsname{0.0}
\expandafter\def\csname val@rev.pol.infeas.uiv.p1@0.25.n20\endcsname{0}
\expandafter\def\csname cnt@rev.pol.infeas.uiv.p1@0.25.n20\endcsname{0}
\expandafter\def\csname pct@rev.pol.infeas.uiv.p1@0.25.n20\endcsname{0.0}
\expandafter\def\csname ppt@rev.pol.infeas.uiv.p1@0.25.n20\endcsname{0.0}
\expandafter\def\csname val@rev.pol.infeas.uiv.p1@0.25.nN\endcsname{0}
\expandafter\def\csname cnt@rev.pol.infeas.uiv.p1@0.25.nN\endcsname{0}
\expandafter\def\csname pct@rev.pol.infeas.uiv.p1@0.25.nN\endcsname{0.0}
\expandafter\def\csname ppt@rev.pol.infeas.uiv.p1@0.25.nN\endcsname{0.0}
\expandafter\def\csname val@rev.pol.infeas.uiv.p1@0.5.n-skills\endcsname{0}
\expandafter\def\csname cnt@rev.pol.infeas.uiv.p1@0.5.n-skills\endcsname{0}
\expandafter\def\csname pct@rev.pol.infeas.uiv.p1@0.5.n-skills\endcsname{0.0}
\expandafter\def\csname ppt@rev.pol.infeas.uiv.p1@0.5.n-skills\endcsname{0.0}
\expandafter\def\csname val@rev.pol.infeas.uiv.p1@0.5.n20\endcsname{0}
\expandafter\def\csname cnt@rev.pol.infeas.uiv.p1@0.5.n20\endcsname{0}
\expandafter\def\csname pct@rev.pol.infeas.uiv.p1@0.5.n20\endcsname{0.0}
\expandafter\def\csname ppt@rev.pol.infeas.uiv.p1@0.5.n20\endcsname{0.0}
\expandafter\def\csname val@rev.pol.infeas.uiv.p1@0.5.nN\endcsname{0}
\expandafter\def\csname cnt@rev.pol.infeas.uiv.p1@0.5.nN\endcsname{0}
\expandafter\def\csname pct@rev.pol.infeas.uiv.p1@0.5.nN\endcsname{0.0}
\expandafter\def\csname ppt@rev.pol.infeas.uiv.p1@0.5.nN\endcsname{0.0}
\expandafter\def\csname val@rev.pol.infeas.uiv.p1@1.0.n-skills\endcsname{0}
\expandafter\def\csname cnt@rev.pol.infeas.uiv.p1@1.0.n-skills\endcsname{0}
\expandafter\def\csname pct@rev.pol.infeas.uiv.p1@1.0.n-skills\endcsname{0.0}
\expandafter\def\csname ppt@rev.pol.infeas.uiv.p1@1.0.n-skills\endcsname{0.0}
\expandafter\def\csname val@rev.pol.infeas.uiv.p1@1.0.n20\endcsname{0}
\expandafter\def\csname cnt@rev.pol.infeas.uiv.p1@1.0.n20\endcsname{0}
\expandafter\def\csname pct@rev.pol.infeas.uiv.p1@1.0.n20\endcsname{0.0}
\expandafter\def\csname ppt@rev.pol.infeas.uiv.p1@1.0.n20\endcsname{0.0}
\expandafter\def\csname val@rev.pol.infeas.uiv.p1@1.0.nN\endcsname{0}
\expandafter\def\csname cnt@rev.pol.infeas.uiv.p1@1.0.nN\endcsname{0}
\expandafter\def\csname pct@rev.pol.infeas.uiv.p1@1.0.nN\endcsname{0.0}
\expandafter\def\csname ppt@rev.pol.infeas.uiv.p1@1.0.nN\endcsname{0.0}
\expandafter\def\csname val@rev.pol.infeas.uiv.p2@0.1.n-skills\endcsname{0}
\expandafter\def\csname cnt@rev.pol.infeas.uiv.p2@0.1.n-skills\endcsname{0}
\expandafter\def\csname pct@rev.pol.infeas.uiv.p2@0.1.n-skills\endcsname{0.0}
\expandafter\def\csname ppt@rev.pol.infeas.uiv.p2@0.1.n-skills\endcsname{0.0}
\expandafter\def\csname val@rev.pol.infeas.uiv.p2@0.1.n20\endcsname{0}
\expandafter\def\csname cnt@rev.pol.infeas.uiv.p2@0.1.n20\endcsname{0}
\expandafter\def\csname pct@rev.pol.infeas.uiv.p2@0.1.n20\endcsname{0.0}
\expandafter\def\csname ppt@rev.pol.infeas.uiv.p2@0.1.n20\endcsname{0.0}
\expandafter\def\csname val@rev.pol.infeas.uiv.p2@0.1.nN\endcsname{0}
\expandafter\def\csname cnt@rev.pol.infeas.uiv.p2@0.1.nN\endcsname{0}
\expandafter\def\csname pct@rev.pol.infeas.uiv.p2@0.1.nN\endcsname{0.0}
\expandafter\def\csname ppt@rev.pol.infeas.uiv.p2@0.1.nN\endcsname{0.0}
\expandafter\def\csname val@rev.pol.infeas.uiv.p2@0.25.n-skills\endcsname{0}
\expandafter\def\csname cnt@rev.pol.infeas.uiv.p2@0.25.n-skills\endcsname{0}
\expandafter\def\csname pct@rev.pol.infeas.uiv.p2@0.25.n-skills\endcsname{0.0}
\expandafter\def\csname ppt@rev.pol.infeas.uiv.p2@0.25.n-skills\endcsname{0.0}
\expandafter\def\csname val@rev.pol.infeas.uiv.p2@0.25.n20\endcsname{0}
\expandafter\def\csname cnt@rev.pol.infeas.uiv.p2@0.25.n20\endcsname{0}
\expandafter\def\csname pct@rev.pol.infeas.uiv.p2@0.25.n20\endcsname{0.0}
\expandafter\def\csname ppt@rev.pol.infeas.uiv.p2@0.25.n20\endcsname{0.0}
\expandafter\def\csname val@rev.pol.infeas.uiv.p2@0.25.nN\endcsname{0}
\expandafter\def\csname cnt@rev.pol.infeas.uiv.p2@0.25.nN\endcsname{0}
\expandafter\def\csname pct@rev.pol.infeas.uiv.p2@0.25.nN\endcsname{0.0}
\expandafter\def\csname ppt@rev.pol.infeas.uiv.p2@0.25.nN\endcsname{0.0}
\expandafter\def\csname val@rev.pol.infeas.uiv.p2@0.5.n-skills\endcsname{0}
\expandafter\def\csname cnt@rev.pol.infeas.uiv.p2@0.5.n-skills\endcsname{0}
\expandafter\def\csname pct@rev.pol.infeas.uiv.p2@0.5.n-skills\endcsname{0.0}
\expandafter\def\csname ppt@rev.pol.infeas.uiv.p2@0.5.n-skills\endcsname{0.0}
\expandafter\def\csname val@rev.pol.infeas.uiv.p2@0.5.n20\endcsname{0}
\expandafter\def\csname cnt@rev.pol.infeas.uiv.p2@0.5.n20\endcsname{0}
\expandafter\def\csname pct@rev.pol.infeas.uiv.p2@0.5.n20\endcsname{0.0}
\expandafter\def\csname ppt@rev.pol.infeas.uiv.p2@0.5.n20\endcsname{0.0}
\expandafter\def\csname val@rev.pol.infeas.uiv.p2@0.5.nN\endcsname{0}
\expandafter\def\csname cnt@rev.pol.infeas.uiv.p2@0.5.nN\endcsname{0}
\expandafter\def\csname pct@rev.pol.infeas.uiv.p2@0.5.nN\endcsname{0.0}
\expandafter\def\csname ppt@rev.pol.infeas.uiv.p2@0.5.nN\endcsname{0.0}
\expandafter\def\csname val@rev.pol.infeas.uiv.p2@1.0.n-skills\endcsname{0}
\expandafter\def\csname cnt@rev.pol.infeas.uiv.p2@1.0.n-skills\endcsname{0}
\expandafter\def\csname pct@rev.pol.infeas.uiv.p2@1.0.n-skills\endcsname{0.0}
\expandafter\def\csname ppt@rev.pol.infeas.uiv.p2@1.0.n-skills\endcsname{0.0}
\expandafter\def\csname val@rev.pol.infeas.uiv.p2@1.0.n20\endcsname{0}
\expandafter\def\csname cnt@rev.pol.infeas.uiv.p2@1.0.n20\endcsname{0}
\expandafter\def\csname pct@rev.pol.infeas.uiv.p2@1.0.n20\endcsname{0.0}
\expandafter\def\csname ppt@rev.pol.infeas.uiv.p2@1.0.n20\endcsname{0.0}
\expandafter\def\csname val@rev.pol.infeas.uiv.p2@1.0.nN\endcsname{0}
\expandafter\def\csname cnt@rev.pol.infeas.uiv.p2@1.0.nN\endcsname{0}
\expandafter\def\csname pct@rev.pol.infeas.uiv.p2@1.0.nN\endcsname{0.0}
\expandafter\def\csname ppt@rev.pol.infeas.uiv.p2@1.0.nN\endcsname{0.0}
\expandafter\def\csname val@rev.pol.keep.sspro.p1.best-single@0.1.share-direct-N\endcsname{0.164}
\expandafter\def\csname cnt@rev.pol.keep.sspro.p1.best-single@0.1.share-direct-N\endcsname{0}
\expandafter\def\csname pct@rev.pol.keep.sspro.p1.best-single@0.1.share-direct-N\endcsname{16.4}
\expandafter\def\csname ppt@rev.pol.keep.sspro.p1.best-single@0.1.share-direct-N\endcsname{$+$16.4}
\expandafter\def\csname val@rev.pol.keep.sspro.p1.best-single@0.25.share-direct-N\endcsname{0.090}
\expandafter\def\csname cnt@rev.pol.keep.sspro.p1.best-single@0.25.share-direct-N\endcsname{0}
\expandafter\def\csname pct@rev.pol.keep.sspro.p1.best-single@0.25.share-direct-N\endcsname{9.0}
\expandafter\def\csname ppt@rev.pol.keep.sspro.p1.best-single@0.25.share-direct-N\endcsname{$+$9.0}
\expandafter\def\csname val@rev.pol.keep.sspro.p1.best-single@0.5.share-direct-N\endcsname{0.006}
\expandafter\def\csname cnt@rev.pol.keep.sspro.p1.best-single@0.5.share-direct-N\endcsname{0}
\expandafter\def\csname pct@rev.pol.keep.sspro.p1.best-single@0.5.share-direct-N\endcsname{0.6}
\expandafter\def\csname ppt@rev.pol.keep.sspro.p1.best-single@0.5.share-direct-N\endcsname{$+$0.6}
\expandafter\def\csname val@rev.pol.keep.sspro.p1.best-single@1.0.share-direct-N\endcsname{0.000}
\expandafter\def\csname cnt@rev.pol.keep.sspro.p1.best-single@1.0.share-direct-N\endcsname{0}
\expandafter\def\csname pct@rev.pol.keep.sspro.p1.best-single@1.0.share-direct-N\endcsname{0.0}
\expandafter\def\csname ppt@rev.pol.keep.sspro.p1.best-single@1.0.share-direct-N\endcsname{0.0}
\expandafter\def\csname val@rev.pol.keep.sspro.p1.corr.budgeted-points\endcsname{16}
\expandafter\def\csname cnt@rev.pol.keep.sspro.p1.corr.budgeted-points\endcsname{16}
\expandafter\def\csname pct@rev.pol.keep.sspro.p1.corr.budgeted-points\endcsname{1600.0}
\expandafter\def\csname ppt@rev.pol.keep.sspro.p1.corr.budgeted-points\endcsname{$+$1600.0}
\expandafter\def\csname val@rev.pol.keep.sspro.p1.corr.pearson\endcsname{1.000}
\expandafter\def\csname cnt@rev.pol.keep.sspro.p1.corr.pearson\endcsname{1}
\expandafter\def\csname pct@rev.pol.keep.sspro.p1.corr.pearson\endcsname{100.0}
\expandafter\def\csname ppt@rev.pol.keep.sspro.p1.corr.pearson\endcsname{$+$100.0}
\expandafter\def\csname val@rev.pol.keep.sspro.p1.corr.pearson-20\endcsname{1.000}
\expandafter\def\csname cnt@rev.pol.keep.sspro.p1.corr.pearson-20\endcsname{1}
\expandafter\def\csname pct@rev.pol.keep.sspro.p1.corr.pearson-20\endcsname{100.0}
\expandafter\def\csname ppt@rev.pol.keep.sspro.p1.corr.pearson-20\endcsname{$+$100.0}
\expandafter\def\csname val@rev.pol.keep.sspro.p1.corr.pearson-all\endcsname{1.000}
\expandafter\def\csname cnt@rev.pol.keep.sspro.p1.corr.pearson-all\endcsname{1}
\expandafter\def\csname pct@rev.pol.keep.sspro.p1.corr.pearson-all\endcsname{100.0}
\expandafter\def\csname ppt@rev.pol.keep.sspro.p1.corr.pearson-all\endcsname{$+$100.0}
\expandafter\def\csname val@rev.pol.keep.sspro.p1.corr.slope-pt-per-keep\endcsname{0.120}
\expandafter\def\csname cnt@rev.pol.keep.sspro.p1.corr.slope-pt-per-keep\endcsname{0}
\expandafter\def\csname pct@rev.pol.keep.sspro.p1.corr.slope-pt-per-keep\endcsname{12.0}
\expandafter\def\csname ppt@rev.pol.keep.sspro.p1.corr.slope-pt-per-keep\endcsname{$+$12.0}
\expandafter\def\csname val@rev.pol.keep.sspro.p1.corr.spearman\endcsname{1.000}
\expandafter\def\csname cnt@rev.pol.keep.sspro.p1.corr.spearman\endcsname{1}
\expandafter\def\csname pct@rev.pol.keep.sspro.p1.corr.spearman\endcsname{100.0}
\expandafter\def\csname ppt@rev.pol.keep.sspro.p1.corr.spearman\endcsname{$+$100.0}
\expandafter\def\csname val@rev.pol.keep.sspro.p1.corr.spearman-20\endcsname{0.993}
\expandafter\def\csname cnt@rev.pol.keep.sspro.p1.corr.spearman-20\endcsname{1}
\expandafter\def\csname pct@rev.pol.keep.sspro.p1.corr.spearman-20\endcsname{99.3}
\expandafter\def\csname ppt@rev.pol.keep.sspro.p1.corr.spearman-20\endcsname{$+$99.3}
\expandafter\def\csname val@rev.pol.keep.sspro.p1.corr.spearman-all\endcsname{1.000}
\expandafter\def\csname cnt@rev.pol.keep.sspro.p1.corr.spearman-all\endcsname{1}
\expandafter\def\csname pct@rev.pol.keep.sspro.p1.corr.spearman-all\endcsname{100.0}
\expandafter\def\csname ppt@rev.pol.keep.sspro.p1.corr.spearman-all\endcsname{$+$100.0}
\expandafter\def\csname val@rev.pol.keep.sspro.p1.corr.spearman-primary\endcsname{1.000}
\expandafter\def\csname cnt@rev.pol.keep.sspro.p1.corr.spearman-primary\endcsname{1}
\expandafter\def\csname pct@rev.pol.keep.sspro.p1.corr.spearman-primary\endcsname{100.0}
\expandafter\def\csname ppt@rev.pol.keep.sspro.p1.corr.spearman-primary\endcsname{$+$100.0}
\expandafter\def\csname val@rev.pol.keep.sspro.p1.corr.with-nobudget-points\endcsname{21}
\expandafter\def\csname cnt@rev.pol.keep.sspro.p1.corr.with-nobudget-points\endcsname{21}
\expandafter\def\csname pct@rev.pol.keep.sspro.p1.corr.with-nobudget-points\endcsname{2100.0}
\expandafter\def\csname ppt@rev.pol.keep.sspro.p1.corr.with-nobudget-points\endcsname{$+$2100.0}
\expandafter\def\csname val@rev.pol.keep.sspro.p1.full-recert.keep20\endcsname{1.000}
\expandafter\def\csname cnt@rev.pol.keep.sspro.p1.full-recert.keep20\endcsname{1}
\expandafter\def\csname pct@rev.pol.keep.sspro.p1.full-recert.keep20\endcsname{100.0}
\expandafter\def\csname ppt@rev.pol.keep.sspro.p1.full-recert.keep20\endcsname{$+$100.0}
\expandafter\def\csname val@rev.pol.keep.sspro.p1.full-recert.keepN\endcsname{1.000}
\expandafter\def\csname cnt@rev.pol.keep.sspro.p1.full-recert.keepN\endcsname{1}
\expandafter\def\csname pct@rev.pol.keep.sspro.p1.full-recert.keepN\endcsname{100.0}
\expandafter\def\csname ppt@rev.pol.keep.sspro.p1.full-recert.keepN\endcsname{$+$100.0}
\expandafter\def\csname val@rev.pol.keep.sspro.p1.full-recert.scoreN\endcsname{0.859}
\expandafter\def\csname cnt@rev.pol.keep.sspro.p1.full-recert.scoreN\endcsname{1}
\expandafter\def\csname pct@rev.pol.keep.sspro.p1.full-recert.scoreN\endcsname{85.9}
\expandafter\def\csname ppt@rev.pol.keep.sspro.p1.full-recert.scoreN\endcsname{$+$85.9}
\expandafter\def\csname val@rev.pol.keep.sspro.p1.generic-ppi-source@0.1.infeasible-rateN\endcsname{0.000}
\expandafter\def\csname cnt@rev.pol.keep.sspro.p1.generic-ppi-source@0.1.infeasible-rateN\endcsname{0}
\expandafter\def\csname pct@rev.pol.keep.sspro.p1.generic-ppi-source@0.1.infeasible-rateN\endcsname{0.0}
\expandafter\def\csname ppt@rev.pol.keep.sspro.p1.generic-ppi-source@0.1.infeasible-rateN\endcsname{0.0}
\expandafter\def\csname val@rev.pol.keep.sspro.p1.generic-ppi-source@0.1.keep20\endcsname{0.912}
\expandafter\def\csname cnt@rev.pol.keep.sspro.p1.generic-ppi-source@0.1.keep20\endcsname{1}
\expandafter\def\csname pct@rev.pol.keep.sspro.p1.generic-ppi-source@0.1.keep20\endcsname{91.2}
\expandafter\def\csname ppt@rev.pol.keep.sspro.p1.generic-ppi-source@0.1.keep20\endcsname{$+$91.2}
\expandafter\def\csname val@rev.pol.keep.sspro.p1.generic-ppi-source@0.1.keepN\endcsname{0.879}
\expandafter\def\csname cnt@rev.pol.keep.sspro.p1.generic-ppi-source@0.1.keepN\endcsname{1}
\expandafter\def\csname pct@rev.pol.keep.sspro.p1.generic-ppi-source@0.1.keepN\endcsname{87.9}
\expandafter\def\csname ppt@rev.pol.keep.sspro.p1.generic-ppi-source@0.1.keepN\endcsname{$+$87.9}
\expandafter\def\csname val@rev.pol.keep.sspro.p1.generic-ppi-source@0.1.scoreN\endcsname{0.845}
\expandafter\def\csname cnt@rev.pol.keep.sspro.p1.generic-ppi-source@0.1.scoreN\endcsname{1}
\expandafter\def\csname pct@rev.pol.keep.sspro.p1.generic-ppi-source@0.1.scoreN\endcsname{84.5}
\expandafter\def\csname ppt@rev.pol.keep.sspro.p1.generic-ppi-source@0.1.scoreN\endcsname{$+$84.5}
\expandafter\def\csname val@rev.pol.keep.sspro.p1.generic-ppi-source@0.25.infeasible-rateN\endcsname{0.000}
\expandafter\def\csname cnt@rev.pol.keep.sspro.p1.generic-ppi-source@0.25.infeasible-rateN\endcsname{0}
\expandafter\def\csname pct@rev.pol.keep.sspro.p1.generic-ppi-source@0.25.infeasible-rateN\endcsname{0.0}
\expandafter\def\csname ppt@rev.pol.keep.sspro.p1.generic-ppi-source@0.25.infeasible-rateN\endcsname{0.0}
\expandafter\def\csname val@rev.pol.keep.sspro.p1.generic-ppi-source@0.25.keep20\endcsname{0.938}
\expandafter\def\csname cnt@rev.pol.keep.sspro.p1.generic-ppi-source@0.25.keep20\endcsname{1}
\expandafter\def\csname pct@rev.pol.keep.sspro.p1.generic-ppi-source@0.25.keep20\endcsname{93.8}
\expandafter\def\csname ppt@rev.pol.keep.sspro.p1.generic-ppi-source@0.25.keep20\endcsname{$+$93.8}
\expandafter\def\csname val@rev.pol.keep.sspro.p1.generic-ppi-source@0.25.keepN\endcsname{0.901}
\expandafter\def\csname cnt@rev.pol.keep.sspro.p1.generic-ppi-source@0.25.keepN\endcsname{1}
\expandafter\def\csname pct@rev.pol.keep.sspro.p1.generic-ppi-source@0.25.keepN\endcsname{90.1}
\expandafter\def\csname ppt@rev.pol.keep.sspro.p1.generic-ppi-source@0.25.keepN\endcsname{$+$90.1}
\expandafter\def\csname val@rev.pol.keep.sspro.p1.generic-ppi-source@0.25.scoreN\endcsname{0.847}
\expandafter\def\csname cnt@rev.pol.keep.sspro.p1.generic-ppi-source@0.25.scoreN\endcsname{1}
\expandafter\def\csname pct@rev.pol.keep.sspro.p1.generic-ppi-source@0.25.scoreN\endcsname{84.7}
\expandafter\def\csname ppt@rev.pol.keep.sspro.p1.generic-ppi-source@0.25.scoreN\endcsname{$+$84.7}
\expandafter\def\csname val@rev.pol.keep.sspro.p1.generic-ppi-source@0.5.infeasible-rateN\endcsname{0.000}
\expandafter\def\csname cnt@rev.pol.keep.sspro.p1.generic-ppi-source@0.5.infeasible-rateN\endcsname{0}
\expandafter\def\csname pct@rev.pol.keep.sspro.p1.generic-ppi-source@0.5.infeasible-rateN\endcsname{0.0}
\expandafter\def\csname ppt@rev.pol.keep.sspro.p1.generic-ppi-source@0.5.infeasible-rateN\endcsname{0.0}
\expandafter\def\csname val@rev.pol.keep.sspro.p1.generic-ppi-source@0.5.keep20\endcsname{1.000}
\expandafter\def\csname cnt@rev.pol.keep.sspro.p1.generic-ppi-source@0.5.keep20\endcsname{1}
\expandafter\def\csname pct@rev.pol.keep.sspro.p1.generic-ppi-source@0.5.keep20\endcsname{100.0}
\expandafter\def\csname ppt@rev.pol.keep.sspro.p1.generic-ppi-source@0.5.keep20\endcsname{$+$100.0}
\expandafter\def\csname val@rev.pol.keep.sspro.p1.generic-ppi-source@0.5.keepN\endcsname{0.990}
\expandafter\def\csname cnt@rev.pol.keep.sspro.p1.generic-ppi-source@0.5.keepN\endcsname{1}
\expandafter\def\csname pct@rev.pol.keep.sspro.p1.generic-ppi-source@0.5.keepN\endcsname{99.0}
\expandafter\def\csname ppt@rev.pol.keep.sspro.p1.generic-ppi-source@0.5.keepN\endcsname{$+$99.0}
\expandafter\def\csname val@rev.pol.keep.sspro.p1.generic-ppi-source@0.5.scoreN\endcsname{0.858}
\expandafter\def\csname cnt@rev.pol.keep.sspro.p1.generic-ppi-source@0.5.scoreN\endcsname{1}
\expandafter\def\csname pct@rev.pol.keep.sspro.p1.generic-ppi-source@0.5.scoreN\endcsname{85.8}
\expandafter\def\csname ppt@rev.pol.keep.sspro.p1.generic-ppi-source@0.5.scoreN\endcsname{$+$85.8}
\expandafter\def\csname val@rev.pol.keep.sspro.p1.generic-ppi-source@1.0.infeasible-rateN\endcsname{0.000}
\expandafter\def\csname cnt@rev.pol.keep.sspro.p1.generic-ppi-source@1.0.infeasible-rateN\endcsname{0}
\expandafter\def\csname pct@rev.pol.keep.sspro.p1.generic-ppi-source@1.0.infeasible-rateN\endcsname{0.0}
\expandafter\def\csname ppt@rev.pol.keep.sspro.p1.generic-ppi-source@1.0.infeasible-rateN\endcsname{0.0}
\expandafter\def\csname val@rev.pol.keep.sspro.p1.generic-ppi-source@1.0.keep20\endcsname{1.000}
\expandafter\def\csname cnt@rev.pol.keep.sspro.p1.generic-ppi-source@1.0.keep20\endcsname{1}
\expandafter\def\csname pct@rev.pol.keep.sspro.p1.generic-ppi-source@1.0.keep20\endcsname{100.0}
\expandafter\def\csname ppt@rev.pol.keep.sspro.p1.generic-ppi-source@1.0.keep20\endcsname{$+$100.0}
\expandafter\def\csname val@rev.pol.keep.sspro.p1.generic-ppi-source@1.0.keepN\endcsname{1.000}
\expandafter\def\csname cnt@rev.pol.keep.sspro.p1.generic-ppi-source@1.0.keepN\endcsname{1}
\expandafter\def\csname pct@rev.pol.keep.sspro.p1.generic-ppi-source@1.0.keepN\endcsname{100.0}
\expandafter\def\csname ppt@rev.pol.keep.sspro.p1.generic-ppi-source@1.0.keepN\endcsname{$+$100.0}
\expandafter\def\csname val@rev.pol.keep.sspro.p1.generic-ppi-source@1.0.scoreN\endcsname{0.859}
\expandafter\def\csname cnt@rev.pol.keep.sspro.p1.generic-ppi-source@1.0.scoreN\endcsname{1}
\expandafter\def\csname pct@rev.pol.keep.sspro.p1.generic-ppi-source@1.0.scoreN\endcsname{85.9}
\expandafter\def\csname ppt@rev.pol.keep.sspro.p1.generic-ppi-source@1.0.scoreN\endcsname{$+$85.9}
\expandafter\def\csname val@rev.pol.keep.sspro.p1.keep-all.keep20\endcsname{1.000}
\expandafter\def\csname cnt@rev.pol.keep.sspro.p1.keep-all.keep20\endcsname{1}
\expandafter\def\csname pct@rev.pol.keep.sspro.p1.keep-all.keep20\endcsname{100.0}
\expandafter\def\csname ppt@rev.pol.keep.sspro.p1.keep-all.keep20\endcsname{$+$100.0}
\expandafter\def\csname val@rev.pol.keep.sspro.p1.keep-all.keepN\endcsname{1.000}
\expandafter\def\csname cnt@rev.pol.keep.sspro.p1.keep-all.keepN\endcsname{1}
\expandafter\def\csname pct@rev.pol.keep.sspro.p1.keep-all.keepN\endcsname{100.0}
\expandafter\def\csname ppt@rev.pol.keep.sspro.p1.keep-all.keepN\endcsname{$+$100.0}
\expandafter\def\csname val@rev.pol.keep.sspro.p1.keep-all.scoreN\endcsname{0.859}
\expandafter\def\csname cnt@rev.pol.keep.sspro.p1.keep-all.scoreN\endcsname{1}
\expandafter\def\csname pct@rev.pol.keep.sspro.p1.keep-all.scoreN\endcsname{85.9}
\expandafter\def\csname ppt@rev.pol.keep.sspro.p1.keep-all.scoreN\endcsname{$+$85.9}
\expandafter\def\csname val@rev.pol.keep.sspro.p1.old-decision.keep20\endcsname{1.000}
\expandafter\def\csname cnt@rev.pol.keep.sspro.p1.old-decision.keep20\endcsname{1}
\expandafter\def\csname pct@rev.pol.keep.sspro.p1.old-decision.keep20\endcsname{100.0}
\expandafter\def\csname ppt@rev.pol.keep.sspro.p1.old-decision.keep20\endcsname{$+$100.0}
\expandafter\def\csname val@rev.pol.keep.sspro.p1.old-decision.keepN\endcsname{1.000}
\expandafter\def\csname cnt@rev.pol.keep.sspro.p1.old-decision.keepN\endcsname{1}
\expandafter\def\csname pct@rev.pol.keep.sspro.p1.old-decision.keepN\endcsname{100.0}
\expandafter\def\csname ppt@rev.pol.keep.sspro.p1.old-decision.keepN\endcsname{$+$100.0}
\expandafter\def\csname val@rev.pol.keep.sspro.p1.old-decision.scoreN\endcsname{0.859}
\expandafter\def\csname cnt@rev.pol.keep.sspro.p1.old-decision.scoreN\endcsname{1}
\expandafter\def\csname pct@rev.pol.keep.sspro.p1.old-decision.scoreN\endcsname{85.9}
\expandafter\def\csname ppt@rev.pol.keep.sspro.p1.old-decision.scoreN\endcsname{$+$85.9}
\expandafter\def\csname val@rev.pol.keep.sspro.p1.oracle-on-T.keep20\endcsname{1.000}
\expandafter\def\csname cnt@rev.pol.keep.sspro.p1.oracle-on-T.keep20\endcsname{1}
\expandafter\def\csname pct@rev.pol.keep.sspro.p1.oracle-on-T.keep20\endcsname{100.0}
\expandafter\def\csname ppt@rev.pol.keep.sspro.p1.oracle-on-T.keep20\endcsname{$+$100.0}
\expandafter\def\csname val@rev.pol.keep.sspro.p1.oracle-on-T.keepN\endcsname{1.000}
\expandafter\def\csname cnt@rev.pol.keep.sspro.p1.oracle-on-T.keepN\endcsname{1}
\expandafter\def\csname pct@rev.pol.keep.sspro.p1.oracle-on-T.keepN\endcsname{100.0}
\expandafter\def\csname ppt@rev.pol.keep.sspro.p1.oracle-on-T.keepN\endcsname{$+$100.0}
\expandafter\def\csname val@rev.pol.keep.sspro.p1.oracle-on-T.scoreN\endcsname{0.859}
\expandafter\def\csname cnt@rev.pol.keep.sspro.p1.oracle-on-T.scoreN\endcsname{1}
\expandafter\def\csname pct@rev.pol.keep.sspro.p1.oracle-on-T.scoreN\endcsname{85.9}
\expandafter\def\csname ppt@rev.pol.keep.sspro.p1.oracle-on-T.scoreN\endcsname{$+$85.9}
\expandafter\def\csname val@rev.pol.keep.sspro.p1.rcr-stratified@0.1.infeasible-rateN\endcsname{1.000}
\expandafter\def\csname cnt@rev.pol.keep.sspro.p1.rcr-stratified@0.1.infeasible-rateN\endcsname{1}
\expandafter\def\csname pct@rev.pol.keep.sspro.p1.rcr-stratified@0.1.infeasible-rateN\endcsname{100.0}
\expandafter\def\csname ppt@rev.pol.keep.sspro.p1.rcr-stratified@0.1.infeasible-rateN\endcsname{$+$100.0}
\expandafter\def\csname val@rev.pol.keep.sspro.p1.rcr-stratified@0.1.keep20\endcsname{0.000}
\expandafter\def\csname cnt@rev.pol.keep.sspro.p1.rcr-stratified@0.1.keep20\endcsname{0}
\expandafter\def\csname pct@rev.pol.keep.sspro.p1.rcr-stratified@0.1.keep20\endcsname{0.0}
\expandafter\def\csname ppt@rev.pol.keep.sspro.p1.rcr-stratified@0.1.keep20\endcsname{0.0}
\expandafter\def\csname val@rev.pol.keep.sspro.p1.rcr-stratified@0.1.keepN\endcsname{0.000}
\expandafter\def\csname cnt@rev.pol.keep.sspro.p1.rcr-stratified@0.1.keepN\endcsname{0}
\expandafter\def\csname pct@rev.pol.keep.sspro.p1.rcr-stratified@0.1.keepN\endcsname{0.0}
\expandafter\def\csname ppt@rev.pol.keep.sspro.p1.rcr-stratified@0.1.keepN\endcsname{0.0}
\expandafter\def\csname val@rev.pol.keep.sspro.p1.rcr-stratified@0.1.scoreN\endcsname{0.740}
\expandafter\def\csname cnt@rev.pol.keep.sspro.p1.rcr-stratified@0.1.scoreN\endcsname{1}
\expandafter\def\csname pct@rev.pol.keep.sspro.p1.rcr-stratified@0.1.scoreN\endcsname{74.0}
\expandafter\def\csname ppt@rev.pol.keep.sspro.p1.rcr-stratified@0.1.scoreN\endcsname{$+$74.0}
\expandafter\def\csname val@rev.pol.keep.sspro.p1.rcr-stratified@0.25.infeasible-rateN\endcsname{0.000}
\expandafter\def\csname cnt@rev.pol.keep.sspro.p1.rcr-stratified@0.25.infeasible-rateN\endcsname{0}
\expandafter\def\csname pct@rev.pol.keep.sspro.p1.rcr-stratified@0.25.infeasible-rateN\endcsname{0.0}
\expandafter\def\csname ppt@rev.pol.keep.sspro.p1.rcr-stratified@0.25.infeasible-rateN\endcsname{0.0}
\expandafter\def\csname val@rev.pol.keep.sspro.p1.rcr-stratified@0.25.keep20\endcsname{0.988}
\expandafter\def\csname cnt@rev.pol.keep.sspro.p1.rcr-stratified@0.25.keep20\endcsname{1}
\expandafter\def\csname pct@rev.pol.keep.sspro.p1.rcr-stratified@0.25.keep20\endcsname{98.8}
\expandafter\def\csname ppt@rev.pol.keep.sspro.p1.rcr-stratified@0.25.keep20\endcsname{$+$98.8}
\expandafter\def\csname val@rev.pol.keep.sspro.p1.rcr-stratified@0.25.keepN\endcsname{0.973}
\expandafter\def\csname cnt@rev.pol.keep.sspro.p1.rcr-stratified@0.25.keepN\endcsname{1}
\expandafter\def\csname pct@rev.pol.keep.sspro.p1.rcr-stratified@0.25.keepN\endcsname{97.3}
\expandafter\def\csname ppt@rev.pol.keep.sspro.p1.rcr-stratified@0.25.keepN\endcsname{$+$97.3}
\expandafter\def\csname val@rev.pol.keep.sspro.p1.rcr-stratified@0.25.scoreN\endcsname{0.856}
\expandafter\def\csname cnt@rev.pol.keep.sspro.p1.rcr-stratified@0.25.scoreN\endcsname{1}
\expandafter\def\csname pct@rev.pol.keep.sspro.p1.rcr-stratified@0.25.scoreN\endcsname{85.6}
\expandafter\def\csname ppt@rev.pol.keep.sspro.p1.rcr-stratified@0.25.scoreN\endcsname{$+$85.6}
\expandafter\def\csname val@rev.pol.keep.sspro.p1.rcr-stratified@0.5.infeasible-rateN\endcsname{0.000}
\expandafter\def\csname cnt@rev.pol.keep.sspro.p1.rcr-stratified@0.5.infeasible-rateN\endcsname{0}
\expandafter\def\csname pct@rev.pol.keep.sspro.p1.rcr-stratified@0.5.infeasible-rateN\endcsname{0.0}
\expandafter\def\csname ppt@rev.pol.keep.sspro.p1.rcr-stratified@0.5.infeasible-rateN\endcsname{0.0}
\expandafter\def\csname val@rev.pol.keep.sspro.p1.rcr-stratified@0.5.keep20\endcsname{1.000}
\expandafter\def\csname cnt@rev.pol.keep.sspro.p1.rcr-stratified@0.5.keep20\endcsname{1}
\expandafter\def\csname pct@rev.pol.keep.sspro.p1.rcr-stratified@0.5.keep20\endcsname{100.0}
\expandafter\def\csname ppt@rev.pol.keep.sspro.p1.rcr-stratified@0.5.keep20\endcsname{$+$100.0}
\expandafter\def\csname val@rev.pol.keep.sspro.p1.rcr-stratified@0.5.keepN\endcsname{0.997}
\expandafter\def\csname cnt@rev.pol.keep.sspro.p1.rcr-stratified@0.5.keepN\endcsname{1}
\expandafter\def\csname pct@rev.pol.keep.sspro.p1.rcr-stratified@0.5.keepN\endcsname{99.7}
\expandafter\def\csname ppt@rev.pol.keep.sspro.p1.rcr-stratified@0.5.keepN\endcsname{$+$99.7}
\expandafter\def\csname val@rev.pol.keep.sspro.p1.rcr-stratified@0.5.scoreN\endcsname{0.859}
\expandafter\def\csname cnt@rev.pol.keep.sspro.p1.rcr-stratified@0.5.scoreN\endcsname{1}
\expandafter\def\csname pct@rev.pol.keep.sspro.p1.rcr-stratified@0.5.scoreN\endcsname{85.9}
\expandafter\def\csname ppt@rev.pol.keep.sspro.p1.rcr-stratified@0.5.scoreN\endcsname{$+$85.9}
\expandafter\def\csname val@rev.pol.keep.sspro.p1.rcr-stratified@1.0.infeasible-rateN\endcsname{0.000}
\expandafter\def\csname cnt@rev.pol.keep.sspro.p1.rcr-stratified@1.0.infeasible-rateN\endcsname{0}
\expandafter\def\csname pct@rev.pol.keep.sspro.p1.rcr-stratified@1.0.infeasible-rateN\endcsname{0.0}
\expandafter\def\csname ppt@rev.pol.keep.sspro.p1.rcr-stratified@1.0.infeasible-rateN\endcsname{0.0}
\expandafter\def\csname val@rev.pol.keep.sspro.p1.rcr-stratified@1.0.keep20\endcsname{1.000}
\expandafter\def\csname cnt@rev.pol.keep.sspro.p1.rcr-stratified@1.0.keep20\endcsname{1}
\expandafter\def\csname pct@rev.pol.keep.sspro.p1.rcr-stratified@1.0.keep20\endcsname{100.0}
\expandafter\def\csname ppt@rev.pol.keep.sspro.p1.rcr-stratified@1.0.keep20\endcsname{$+$100.0}
\expandafter\def\csname val@rev.pol.keep.sspro.p1.rcr-stratified@1.0.keepN\endcsname{1.000}
\expandafter\def\csname cnt@rev.pol.keep.sspro.p1.rcr-stratified@1.0.keepN\endcsname{1}
\expandafter\def\csname pct@rev.pol.keep.sspro.p1.rcr-stratified@1.0.keepN\endcsname{100.0}
\expandafter\def\csname ppt@rev.pol.keep.sspro.p1.rcr-stratified@1.0.keepN\endcsname{$+$100.0}
\expandafter\def\csname val@rev.pol.keep.sspro.p1.rcr-stratified@1.0.scoreN\endcsname{0.859}
\expandafter\def\csname cnt@rev.pol.keep.sspro.p1.rcr-stratified@1.0.scoreN\endcsname{1}
\expandafter\def\csname pct@rev.pol.keep.sspro.p1.rcr-stratified@1.0.scoreN\endcsname{85.9}
\expandafter\def\csname ppt@rev.pol.keep.sspro.p1.rcr-stratified@1.0.scoreN\endcsname{$+$85.9}
\expandafter\def\csname val@rev.pol.keep.sspro.p1.replay-only.keep20\endcsname{1.000}
\expandafter\def\csname cnt@rev.pol.keep.sspro.p1.replay-only.keep20\endcsname{1}
\expandafter\def\csname pct@rev.pol.keep.sspro.p1.replay-only.keep20\endcsname{100.0}
\expandafter\def\csname ppt@rev.pol.keep.sspro.p1.replay-only.keep20\endcsname{$+$100.0}
\expandafter\def\csname val@rev.pol.keep.sspro.p1.replay-only.keepN\endcsname{1.000}
\expandafter\def\csname cnt@rev.pol.keep.sspro.p1.replay-only.keepN\endcsname{1}
\expandafter\def\csname pct@rev.pol.keep.sspro.p1.replay-only.keepN\endcsname{100.0}
\expandafter\def\csname ppt@rev.pol.keep.sspro.p1.replay-only.keepN\endcsname{$+$100.0}
\expandafter\def\csname val@rev.pol.keep.sspro.p1.replay-only.scoreN\endcsname{0.859}
\expandafter\def\csname cnt@rev.pol.keep.sspro.p1.replay-only.scoreN\endcsname{1}
\expandafter\def\csname pct@rev.pol.keep.sspro.p1.replay-only.scoreN\endcsname{85.9}
\expandafter\def\csname ppt@rev.pol.keep.sspro.p1.replay-only.scoreN\endcsname{$+$85.9}
\expandafter\def\csname val@rev.pol.keep.sspro.p1.replay-ppi-uniform@0.1.infeasible-rateN\endcsname{0.000}
\expandafter\def\csname cnt@rev.pol.keep.sspro.p1.replay-ppi-uniform@0.1.infeasible-rateN\endcsname{0}
\expandafter\def\csname pct@rev.pol.keep.sspro.p1.replay-ppi-uniform@0.1.infeasible-rateN\endcsname{0.0}
\expandafter\def\csname ppt@rev.pol.keep.sspro.p1.replay-ppi-uniform@0.1.infeasible-rateN\endcsname{0.0}
\expandafter\def\csname val@rev.pol.keep.sspro.p1.replay-ppi-uniform@0.1.keep20\endcsname{0.900}
\expandafter\def\csname cnt@rev.pol.keep.sspro.p1.replay-ppi-uniform@0.1.keep20\endcsname{1}
\expandafter\def\csname pct@rev.pol.keep.sspro.p1.replay-ppi-uniform@0.1.keep20\endcsname{90.0}
\expandafter\def\csname ppt@rev.pol.keep.sspro.p1.replay-ppi-uniform@0.1.keep20\endcsname{$+$90.0}
\expandafter\def\csname val@rev.pol.keep.sspro.p1.replay-ppi-uniform@0.1.keepN\endcsname{0.865}
\expandafter\def\csname cnt@rev.pol.keep.sspro.p1.replay-ppi-uniform@0.1.keepN\endcsname{1}
\expandafter\def\csname pct@rev.pol.keep.sspro.p1.replay-ppi-uniform@0.1.keepN\endcsname{86.5}
\expandafter\def\csname ppt@rev.pol.keep.sspro.p1.replay-ppi-uniform@0.1.keepN\endcsname{$+$86.5}
\expandafter\def\csname val@rev.pol.keep.sspro.p1.replay-ppi-uniform@0.1.scoreN\endcsname{0.843}
\expandafter\def\csname cnt@rev.pol.keep.sspro.p1.replay-ppi-uniform@0.1.scoreN\endcsname{1}
\expandafter\def\csname pct@rev.pol.keep.sspro.p1.replay-ppi-uniform@0.1.scoreN\endcsname{84.3}
\expandafter\def\csname ppt@rev.pol.keep.sspro.p1.replay-ppi-uniform@0.1.scoreN\endcsname{$+$84.3}
\expandafter\def\csname val@rev.pol.keep.sspro.p1.replay-ppi-uniform@0.25.infeasible-rateN\endcsname{0.000}
\expandafter\def\csname cnt@rev.pol.keep.sspro.p1.replay-ppi-uniform@0.25.infeasible-rateN\endcsname{0}
\expandafter\def\csname pct@rev.pol.keep.sspro.p1.replay-ppi-uniform@0.25.infeasible-rateN\endcsname{0.0}
\expandafter\def\csname ppt@rev.pol.keep.sspro.p1.replay-ppi-uniform@0.25.infeasible-rateN\endcsname{0.0}
\expandafter\def\csname val@rev.pol.keep.sspro.p1.replay-ppi-uniform@0.25.keep20\endcsname{0.925}
\expandafter\def\csname cnt@rev.pol.keep.sspro.p1.replay-ppi-uniform@0.25.keep20\endcsname{1}
\expandafter\def\csname pct@rev.pol.keep.sspro.p1.replay-ppi-uniform@0.25.keep20\endcsname{92.5}
\expandafter\def\csname ppt@rev.pol.keep.sspro.p1.replay-ppi-uniform@0.25.keep20\endcsname{$+$92.5}
\expandafter\def\csname val@rev.pol.keep.sspro.p1.replay-ppi-uniform@0.25.keepN\endcsname{0.949}
\expandafter\def\csname cnt@rev.pol.keep.sspro.p1.replay-ppi-uniform@0.25.keepN\endcsname{1}
\expandafter\def\csname pct@rev.pol.keep.sspro.p1.replay-ppi-uniform@0.25.keepN\endcsname{94.9}
\expandafter\def\csname ppt@rev.pol.keep.sspro.p1.replay-ppi-uniform@0.25.keepN\endcsname{$+$94.9}
\expandafter\def\csname val@rev.pol.keep.sspro.p1.replay-ppi-uniform@0.25.scoreN\endcsname{0.852}
\expandafter\def\csname cnt@rev.pol.keep.sspro.p1.replay-ppi-uniform@0.25.scoreN\endcsname{1}
\expandafter\def\csname pct@rev.pol.keep.sspro.p1.replay-ppi-uniform@0.25.scoreN\endcsname{85.2}
\expandafter\def\csname ppt@rev.pol.keep.sspro.p1.replay-ppi-uniform@0.25.scoreN\endcsname{$+$85.2}
\expandafter\def\csname val@rev.pol.keep.sspro.p1.replay-ppi-uniform@0.5.infeasible-rateN\endcsname{0.000}
\expandafter\def\csname cnt@rev.pol.keep.sspro.p1.replay-ppi-uniform@0.5.infeasible-rateN\endcsname{0}
\expandafter\def\csname pct@rev.pol.keep.sspro.p1.replay-ppi-uniform@0.5.infeasible-rateN\endcsname{0.0}
\expandafter\def\csname ppt@rev.pol.keep.sspro.p1.replay-ppi-uniform@0.5.infeasible-rateN\endcsname{0.0}
\expandafter\def\csname val@rev.pol.keep.sspro.p1.replay-ppi-uniform@0.5.keep20\endcsname{1.000}
\expandafter\def\csname cnt@rev.pol.keep.sspro.p1.replay-ppi-uniform@0.5.keep20\endcsname{1}
\expandafter\def\csname pct@rev.pol.keep.sspro.p1.replay-ppi-uniform@0.5.keep20\endcsname{100.0}
\expandafter\def\csname ppt@rev.pol.keep.sspro.p1.replay-ppi-uniform@0.5.keep20\endcsname{$+$100.0}
\expandafter\def\csname val@rev.pol.keep.sspro.p1.replay-ppi-uniform@0.5.keepN\endcsname{0.998}
\expandafter\def\csname cnt@rev.pol.keep.sspro.p1.replay-ppi-uniform@0.5.keepN\endcsname{1}
\expandafter\def\csname pct@rev.pol.keep.sspro.p1.replay-ppi-uniform@0.5.keepN\endcsname{99.8}
\expandafter\def\csname ppt@rev.pol.keep.sspro.p1.replay-ppi-uniform@0.5.keepN\endcsname{$+$99.8}
\expandafter\def\csname val@rev.pol.keep.sspro.p1.replay-ppi-uniform@0.5.scoreN\endcsname{0.859}
\expandafter\def\csname cnt@rev.pol.keep.sspro.p1.replay-ppi-uniform@0.5.scoreN\endcsname{1}
\expandafter\def\csname pct@rev.pol.keep.sspro.p1.replay-ppi-uniform@0.5.scoreN\endcsname{85.9}
\expandafter\def\csname ppt@rev.pol.keep.sspro.p1.replay-ppi-uniform@0.5.scoreN\endcsname{$+$85.9}
\expandafter\def\csname val@rev.pol.keep.sspro.p1.replay-ppi-uniform@1.0.infeasible-rateN\endcsname{0.000}
\expandafter\def\csname cnt@rev.pol.keep.sspro.p1.replay-ppi-uniform@1.0.infeasible-rateN\endcsname{0}
\expandafter\def\csname pct@rev.pol.keep.sspro.p1.replay-ppi-uniform@1.0.infeasible-rateN\endcsname{0.0}
\expandafter\def\csname ppt@rev.pol.keep.sspro.p1.replay-ppi-uniform@1.0.infeasible-rateN\endcsname{0.0}
\expandafter\def\csname val@rev.pol.keep.sspro.p1.replay-ppi-uniform@1.0.keep20\endcsname{1.000}
\expandafter\def\csname cnt@rev.pol.keep.sspro.p1.replay-ppi-uniform@1.0.keep20\endcsname{1}
\expandafter\def\csname pct@rev.pol.keep.sspro.p1.replay-ppi-uniform@1.0.keep20\endcsname{100.0}
\expandafter\def\csname ppt@rev.pol.keep.sspro.p1.replay-ppi-uniform@1.0.keep20\endcsname{$+$100.0}
\expandafter\def\csname val@rev.pol.keep.sspro.p1.replay-ppi-uniform@1.0.keepN\endcsname{1.000}
\expandafter\def\csname cnt@rev.pol.keep.sspro.p1.replay-ppi-uniform@1.0.keepN\endcsname{1}
\expandafter\def\csname pct@rev.pol.keep.sspro.p1.replay-ppi-uniform@1.0.keepN\endcsname{100.0}
\expandafter\def\csname ppt@rev.pol.keep.sspro.p1.replay-ppi-uniform@1.0.keepN\endcsname{$+$100.0}
\expandafter\def\csname val@rev.pol.keep.sspro.p1.replay-ppi-uniform@1.0.scoreN\endcsname{0.859}
\expandafter\def\csname cnt@rev.pol.keep.sspro.p1.replay-ppi-uniform@1.0.scoreN\endcsname{1}
\expandafter\def\csname pct@rev.pol.keep.sspro.p1.replay-ppi-uniform@1.0.scoreN\endcsname{85.9}
\expandafter\def\csname ppt@rev.pol.keep.sspro.p1.replay-ppi-uniform@1.0.scoreN\endcsname{$+$85.9}
\expandafter\def\csname val@rev.pol.keep.sspro.p1.suspend-all.keep20\endcsname{0.000}
\expandafter\def\csname cnt@rev.pol.keep.sspro.p1.suspend-all.keep20\endcsname{0}
\expandafter\def\csname pct@rev.pol.keep.sspro.p1.suspend-all.keep20\endcsname{0.0}
\expandafter\def\csname ppt@rev.pol.keep.sspro.p1.suspend-all.keep20\endcsname{0.0}
\expandafter\def\csname val@rev.pol.keep.sspro.p1.suspend-all.keepN\endcsname{0.000}
\expandafter\def\csname cnt@rev.pol.keep.sspro.p1.suspend-all.keepN\endcsname{0}
\expandafter\def\csname pct@rev.pol.keep.sspro.p1.suspend-all.keepN\endcsname{0.0}
\expandafter\def\csname ppt@rev.pol.keep.sspro.p1.suspend-all.keepN\endcsname{0.0}
\expandafter\def\csname val@rev.pol.keep.sspro.p1.suspend-all.scoreN\endcsname{0.740}
\expandafter\def\csname cnt@rev.pol.keep.sspro.p1.suspend-all.scoreN\endcsname{1}
\expandafter\def\csname pct@rev.pol.keep.sspro.p1.suspend-all.scoreN\endcsname{74.0}
\expandafter\def\csname ppt@rev.pol.keep.sspro.p1.suspend-all.scoreN\endcsname{$+$74.0}
\expandafter\def\csname val@rev.pol.keep.sspro.p1.uniform-fresh@0.1.infeasible-rateN\endcsname{0.000}
\expandafter\def\csname cnt@rev.pol.keep.sspro.p1.uniform-fresh@0.1.infeasible-rateN\endcsname{0}
\expandafter\def\csname pct@rev.pol.keep.sspro.p1.uniform-fresh@0.1.infeasible-rateN\endcsname{0.0}
\expandafter\def\csname ppt@rev.pol.keep.sspro.p1.uniform-fresh@0.1.infeasible-rateN\endcsname{0.0}
\expandafter\def\csname val@rev.pol.keep.sspro.p1.uniform-fresh@0.1.keep20\endcsname{0.900}
\expandafter\def\csname cnt@rev.pol.keep.sspro.p1.uniform-fresh@0.1.keep20\endcsname{1}
\expandafter\def\csname pct@rev.pol.keep.sspro.p1.uniform-fresh@0.1.keep20\endcsname{90.0}
\expandafter\def\csname ppt@rev.pol.keep.sspro.p1.uniform-fresh@0.1.keep20\endcsname{$+$90.0}
\expandafter\def\csname val@rev.pol.keep.sspro.p1.uniform-fresh@0.1.keepN\endcsname{0.916}
\expandafter\def\csname cnt@rev.pol.keep.sspro.p1.uniform-fresh@0.1.keepN\endcsname{1}
\expandafter\def\csname pct@rev.pol.keep.sspro.p1.uniform-fresh@0.1.keepN\endcsname{91.6}
\expandafter\def\csname ppt@rev.pol.keep.sspro.p1.uniform-fresh@0.1.keepN\endcsname{$+$91.6}
\expandafter\def\csname val@rev.pol.keep.sspro.p1.uniform-fresh@0.1.scoreN\endcsname{0.849}
\expandafter\def\csname cnt@rev.pol.keep.sspro.p1.uniform-fresh@0.1.scoreN\endcsname{1}
\expandafter\def\csname pct@rev.pol.keep.sspro.p1.uniform-fresh@0.1.scoreN\endcsname{84.9}
\expandafter\def\csname ppt@rev.pol.keep.sspro.p1.uniform-fresh@0.1.scoreN\endcsname{$+$84.9}
\expandafter\def\csname val@rev.pol.keep.sspro.p1.uniform-fresh@0.25.infeasible-rateN\endcsname{0.000}
\expandafter\def\csname cnt@rev.pol.keep.sspro.p1.uniform-fresh@0.25.infeasible-rateN\endcsname{0}
\expandafter\def\csname pct@rev.pol.keep.sspro.p1.uniform-fresh@0.25.infeasible-rateN\endcsname{0.0}
\expandafter\def\csname ppt@rev.pol.keep.sspro.p1.uniform-fresh@0.25.infeasible-rateN\endcsname{0.0}
\expandafter\def\csname val@rev.pol.keep.sspro.p1.uniform-fresh@0.25.keep20\endcsname{0.925}
\expandafter\def\csname cnt@rev.pol.keep.sspro.p1.uniform-fresh@0.25.keep20\endcsname{1}
\expandafter\def\csname pct@rev.pol.keep.sspro.p1.uniform-fresh@0.25.keep20\endcsname{92.5}
\expandafter\def\csname ppt@rev.pol.keep.sspro.p1.uniform-fresh@0.25.keep20\endcsname{$+$92.5}
\expandafter\def\csname val@rev.pol.keep.sspro.p1.uniform-fresh@0.25.keepN\endcsname{0.949}
\expandafter\def\csname cnt@rev.pol.keep.sspro.p1.uniform-fresh@0.25.keepN\endcsname{1}
\expandafter\def\csname pct@rev.pol.keep.sspro.p1.uniform-fresh@0.25.keepN\endcsname{94.9}
\expandafter\def\csname ppt@rev.pol.keep.sspro.p1.uniform-fresh@0.25.keepN\endcsname{$+$94.9}
\expandafter\def\csname val@rev.pol.keep.sspro.p1.uniform-fresh@0.25.scoreN\endcsname{0.853}
\expandafter\def\csname cnt@rev.pol.keep.sspro.p1.uniform-fresh@0.25.scoreN\endcsname{1}
\expandafter\def\csname pct@rev.pol.keep.sspro.p1.uniform-fresh@0.25.scoreN\endcsname{85.3}
\expandafter\def\csname ppt@rev.pol.keep.sspro.p1.uniform-fresh@0.25.scoreN\endcsname{$+$85.3}
\expandafter\def\csname val@rev.pol.keep.sspro.p1.uniform-fresh@0.5.infeasible-rateN\endcsname{0.000}
\expandafter\def\csname cnt@rev.pol.keep.sspro.p1.uniform-fresh@0.5.infeasible-rateN\endcsname{0}
\expandafter\def\csname pct@rev.pol.keep.sspro.p1.uniform-fresh@0.5.infeasible-rateN\endcsname{0.0}
\expandafter\def\csname ppt@rev.pol.keep.sspro.p1.uniform-fresh@0.5.infeasible-rateN\endcsname{0.0}
\expandafter\def\csname val@rev.pol.keep.sspro.p1.uniform-fresh@0.5.keep20\endcsname{1.000}
\expandafter\def\csname cnt@rev.pol.keep.sspro.p1.uniform-fresh@0.5.keep20\endcsname{1}
\expandafter\def\csname pct@rev.pol.keep.sspro.p1.uniform-fresh@0.5.keep20\endcsname{100.0}
\expandafter\def\csname ppt@rev.pol.keep.sspro.p1.uniform-fresh@0.5.keep20\endcsname{$+$100.0}
\expandafter\def\csname val@rev.pol.keep.sspro.p1.uniform-fresh@0.5.keepN\endcsname{0.992}
\expandafter\def\csname cnt@rev.pol.keep.sspro.p1.uniform-fresh@0.5.keepN\endcsname{1}
\expandafter\def\csname pct@rev.pol.keep.sspro.p1.uniform-fresh@0.5.keepN\endcsname{99.2}
\expandafter\def\csname ppt@rev.pol.keep.sspro.p1.uniform-fresh@0.5.keepN\endcsname{$+$99.2}
\expandafter\def\csname val@rev.pol.keep.sspro.p1.uniform-fresh@0.5.scoreN\endcsname{0.858}
\expandafter\def\csname cnt@rev.pol.keep.sspro.p1.uniform-fresh@0.5.scoreN\endcsname{1}
\expandafter\def\csname pct@rev.pol.keep.sspro.p1.uniform-fresh@0.5.scoreN\endcsname{85.8}
\expandafter\def\csname ppt@rev.pol.keep.sspro.p1.uniform-fresh@0.5.scoreN\endcsname{$+$85.8}
\expandafter\def\csname val@rev.pol.keep.sspro.p1.uniform-fresh@1.0.infeasible-rateN\endcsname{0.000}
\expandafter\def\csname cnt@rev.pol.keep.sspro.p1.uniform-fresh@1.0.infeasible-rateN\endcsname{0}
\expandafter\def\csname pct@rev.pol.keep.sspro.p1.uniform-fresh@1.0.infeasible-rateN\endcsname{0.0}
\expandafter\def\csname ppt@rev.pol.keep.sspro.p1.uniform-fresh@1.0.infeasible-rateN\endcsname{0.0}
\expandafter\def\csname val@rev.pol.keep.sspro.p1.uniform-fresh@1.0.keep20\endcsname{1.000}
\expandafter\def\csname cnt@rev.pol.keep.sspro.p1.uniform-fresh@1.0.keep20\endcsname{1}
\expandafter\def\csname pct@rev.pol.keep.sspro.p1.uniform-fresh@1.0.keep20\endcsname{100.0}
\expandafter\def\csname ppt@rev.pol.keep.sspro.p1.uniform-fresh@1.0.keep20\endcsname{$+$100.0}
\expandafter\def\csname val@rev.pol.keep.sspro.p1.uniform-fresh@1.0.keepN\endcsname{1.000}
\expandafter\def\csname cnt@rev.pol.keep.sspro.p1.uniform-fresh@1.0.keepN\endcsname{1}
\expandafter\def\csname pct@rev.pol.keep.sspro.p1.uniform-fresh@1.0.keepN\endcsname{100.0}
\expandafter\def\csname ppt@rev.pol.keep.sspro.p1.uniform-fresh@1.0.keepN\endcsname{$+$100.0}
\expandafter\def\csname val@rev.pol.keep.sspro.p1.uniform-fresh@1.0.scoreN\endcsname{0.859}
\expandafter\def\csname cnt@rev.pol.keep.sspro.p1.uniform-fresh@1.0.scoreN\endcsname{1}
\expandafter\def\csname pct@rev.pol.keep.sspro.p1.uniform-fresh@1.0.scoreN\endcsname{85.9}
\expandafter\def\csname ppt@rev.pol.keep.sspro.p1.uniform-fresh@1.0.scoreN\endcsname{$+$85.9}
\expandafter\def\csname val@rev.pol.keep.sspro.p2.best-single@0.1.share-direct-N\endcsname{0.075}
\expandafter\def\csname cnt@rev.pol.keep.sspro.p2.best-single@0.1.share-direct-N\endcsname{0}
\expandafter\def\csname pct@rev.pol.keep.sspro.p2.best-single@0.1.share-direct-N\endcsname{7.5}
\expandafter\def\csname ppt@rev.pol.keep.sspro.p2.best-single@0.1.share-direct-N\endcsname{$+$7.5}
\expandafter\def\csname val@rev.pol.keep.sspro.p2.best-single@0.25.share-direct-N\endcsname{0.026}
\expandafter\def\csname cnt@rev.pol.keep.sspro.p2.best-single@0.25.share-direct-N\endcsname{0}
\expandafter\def\csname pct@rev.pol.keep.sspro.p2.best-single@0.25.share-direct-N\endcsname{2.6}
\expandafter\def\csname ppt@rev.pol.keep.sspro.p2.best-single@0.25.share-direct-N\endcsname{$+$2.6}
\expandafter\def\csname val@rev.pol.keep.sspro.p2.best-single@0.5.share-direct-N\endcsname{0.000}
\expandafter\def\csname cnt@rev.pol.keep.sspro.p2.best-single@0.5.share-direct-N\endcsname{0}
\expandafter\def\csname pct@rev.pol.keep.sspro.p2.best-single@0.5.share-direct-N\endcsname{0.0}
\expandafter\def\csname ppt@rev.pol.keep.sspro.p2.best-single@0.5.share-direct-N\endcsname{0.0}
\expandafter\def\csname val@rev.pol.keep.sspro.p2.best-single@1.0.share-direct-N\endcsname{0.000}
\expandafter\def\csname cnt@rev.pol.keep.sspro.p2.best-single@1.0.share-direct-N\endcsname{0}
\expandafter\def\csname pct@rev.pol.keep.sspro.p2.best-single@1.0.share-direct-N\endcsname{0.0}
\expandafter\def\csname ppt@rev.pol.keep.sspro.p2.best-single@1.0.share-direct-N\endcsname{0.0}
\expandafter\def\csname val@rev.pol.keep.sspro.p2.corr.budgeted-points\endcsname{16}
\expandafter\def\csname cnt@rev.pol.keep.sspro.p2.corr.budgeted-points\endcsname{16}
\expandafter\def\csname pct@rev.pol.keep.sspro.p2.corr.budgeted-points\endcsname{1600.0}
\expandafter\def\csname ppt@rev.pol.keep.sspro.p2.corr.budgeted-points\endcsname{$+$1600.0}
\expandafter\def\csname val@rev.pol.keep.sspro.p2.corr.pearson\endcsname{1.000}
\expandafter\def\csname cnt@rev.pol.keep.sspro.p2.corr.pearson\endcsname{1}
\expandafter\def\csname pct@rev.pol.keep.sspro.p2.corr.pearson\endcsname{100.0}
\expandafter\def\csname ppt@rev.pol.keep.sspro.p2.corr.pearson\endcsname{$+$100.0}
\expandafter\def\csname val@rev.pol.keep.sspro.p2.corr.pearson-20\endcsname{1.000}
\expandafter\def\csname cnt@rev.pol.keep.sspro.p2.corr.pearson-20\endcsname{1}
\expandafter\def\csname pct@rev.pol.keep.sspro.p2.corr.pearson-20\endcsname{100.0}
\expandafter\def\csname ppt@rev.pol.keep.sspro.p2.corr.pearson-20\endcsname{$+$100.0}
\expandafter\def\csname val@rev.pol.keep.sspro.p2.corr.pearson-all\endcsname{1.000}
\expandafter\def\csname cnt@rev.pol.keep.sspro.p2.corr.pearson-all\endcsname{1}
\expandafter\def\csname pct@rev.pol.keep.sspro.p2.corr.pearson-all\endcsname{100.0}
\expandafter\def\csname ppt@rev.pol.keep.sspro.p2.corr.pearson-all\endcsname{$+$100.0}
\expandafter\def\csname val@rev.pol.keep.sspro.p2.corr.slope-pt-per-keep\endcsname{0.148}
\expandafter\def\csname cnt@rev.pol.keep.sspro.p2.corr.slope-pt-per-keep\endcsname{0}
\expandafter\def\csname pct@rev.pol.keep.sspro.p2.corr.slope-pt-per-keep\endcsname{14.8}
\expandafter\def\csname ppt@rev.pol.keep.sspro.p2.corr.slope-pt-per-keep\endcsname{$+$14.8}
\expandafter\def\csname val@rev.pol.keep.sspro.p2.corr.spearman\endcsname{0.997}
\expandafter\def\csname cnt@rev.pol.keep.sspro.p2.corr.spearman\endcsname{1}
\expandafter\def\csname pct@rev.pol.keep.sspro.p2.corr.spearman\endcsname{99.7}
\expandafter\def\csname ppt@rev.pol.keep.sspro.p2.corr.spearman\endcsname{$+$99.7}
\expandafter\def\csname val@rev.pol.keep.sspro.p2.corr.spearman-20\endcsname{0.997}
\expandafter\def\csname cnt@rev.pol.keep.sspro.p2.corr.spearman-20\endcsname{1}
\expandafter\def\csname pct@rev.pol.keep.sspro.p2.corr.spearman-20\endcsname{99.7}
\expandafter\def\csname ppt@rev.pol.keep.sspro.p2.corr.spearman-20\endcsname{$+$99.7}
\expandafter\def\csname val@rev.pol.keep.sspro.p2.corr.spearman-all\endcsname{0.999}
\expandafter\def\csname cnt@rev.pol.keep.sspro.p2.corr.spearman-all\endcsname{1}
\expandafter\def\csname pct@rev.pol.keep.sspro.p2.corr.spearman-all\endcsname{99.9}
\expandafter\def\csname ppt@rev.pol.keep.sspro.p2.corr.spearman-all\endcsname{$+$99.9}
\expandafter\def\csname val@rev.pol.keep.sspro.p2.corr.spearman-primary\endcsname{0.800}
\expandafter\def\csname cnt@rev.pol.keep.sspro.p2.corr.spearman-primary\endcsname{1}
\expandafter\def\csname pct@rev.pol.keep.sspro.p2.corr.spearman-primary\endcsname{80.0}
\expandafter\def\csname ppt@rev.pol.keep.sspro.p2.corr.spearman-primary\endcsname{$+$80.0}
\expandafter\def\csname val@rev.pol.keep.sspro.p2.corr.with-nobudget-points\endcsname{21}
\expandafter\def\csname cnt@rev.pol.keep.sspro.p2.corr.with-nobudget-points\endcsname{21}
\expandafter\def\csname pct@rev.pol.keep.sspro.p2.corr.with-nobudget-points\endcsname{2100.0}
\expandafter\def\csname ppt@rev.pol.keep.sspro.p2.corr.with-nobudget-points\endcsname{$+$2100.0}
\expandafter\def\csname val@rev.pol.keep.sspro.p2.full-recert.keep20\endcsname{1.000}
\expandafter\def\csname cnt@rev.pol.keep.sspro.p2.full-recert.keep20\endcsname{1}
\expandafter\def\csname pct@rev.pol.keep.sspro.p2.full-recert.keep20\endcsname{100.0}
\expandafter\def\csname ppt@rev.pol.keep.sspro.p2.full-recert.keep20\endcsname{$+$100.0}
\expandafter\def\csname val@rev.pol.keep.sspro.p2.full-recert.keepN\endcsname{1.000}
\expandafter\def\csname cnt@rev.pol.keep.sspro.p2.full-recert.keepN\endcsname{1}
\expandafter\def\csname pct@rev.pol.keep.sspro.p2.full-recert.keepN\endcsname{100.0}
\expandafter\def\csname ppt@rev.pol.keep.sspro.p2.full-recert.keepN\endcsname{$+$100.0}
\expandafter\def\csname val@rev.pol.keep.sspro.p2.full-recert.scoreN\endcsname{0.906}
\expandafter\def\csname cnt@rev.pol.keep.sspro.p2.full-recert.scoreN\endcsname{1}
\expandafter\def\csname pct@rev.pol.keep.sspro.p2.full-recert.scoreN\endcsname{90.6}
\expandafter\def\csname ppt@rev.pol.keep.sspro.p2.full-recert.scoreN\endcsname{$+$90.6}
\expandafter\def\csname val@rev.pol.keep.sspro.p2.generic-ppi-source@0.1.infeasible-rateN\endcsname{0.000}
\expandafter\def\csname cnt@rev.pol.keep.sspro.p2.generic-ppi-source@0.1.infeasible-rateN\endcsname{0}
\expandafter\def\csname pct@rev.pol.keep.sspro.p2.generic-ppi-source@0.1.infeasible-rateN\endcsname{0.0}
\expandafter\def\csname ppt@rev.pol.keep.sspro.p2.generic-ppi-source@0.1.infeasible-rateN\endcsname{0.0}
\expandafter\def\csname val@rev.pol.keep.sspro.p2.generic-ppi-source@0.1.keep20\endcsname{0.925}
\expandafter\def\csname cnt@rev.pol.keep.sspro.p2.generic-ppi-source@0.1.keep20\endcsname{1}
\expandafter\def\csname pct@rev.pol.keep.sspro.p2.generic-ppi-source@0.1.keep20\endcsname{92.5}
\expandafter\def\csname ppt@rev.pol.keep.sspro.p2.generic-ppi-source@0.1.keep20\endcsname{$+$92.5}
\expandafter\def\csname val@rev.pol.keep.sspro.p2.generic-ppi-source@0.1.keepN\endcsname{0.915}
\expandafter\def\csname cnt@rev.pol.keep.sspro.p2.generic-ppi-source@0.1.keepN\endcsname{1}
\expandafter\def\csname pct@rev.pol.keep.sspro.p2.generic-ppi-source@0.1.keepN\endcsname{91.5}
\expandafter\def\csname ppt@rev.pol.keep.sspro.p2.generic-ppi-source@0.1.keepN\endcsname{$+$91.5}
\expandafter\def\csname val@rev.pol.keep.sspro.p2.generic-ppi-source@0.1.scoreN\endcsname{0.893}
\expandafter\def\csname cnt@rev.pol.keep.sspro.p2.generic-ppi-source@0.1.scoreN\endcsname{1}
\expandafter\def\csname pct@rev.pol.keep.sspro.p2.generic-ppi-source@0.1.scoreN\endcsname{89.3}
\expandafter\def\csname ppt@rev.pol.keep.sspro.p2.generic-ppi-source@0.1.scoreN\endcsname{$+$89.3}
\expandafter\def\csname val@rev.pol.keep.sspro.p2.generic-ppi-source@0.25.infeasible-rateN\endcsname{0.000}
\expandafter\def\csname cnt@rev.pol.keep.sspro.p2.generic-ppi-source@0.25.infeasible-rateN\endcsname{0}
\expandafter\def\csname pct@rev.pol.keep.sspro.p2.generic-ppi-source@0.25.infeasible-rateN\endcsname{0.0}
\expandafter\def\csname ppt@rev.pol.keep.sspro.p2.generic-ppi-source@0.25.infeasible-rateN\endcsname{0.0}
\expandafter\def\csname val@rev.pol.keep.sspro.p2.generic-ppi-source@0.25.keep20\endcsname{0.863}
\expandafter\def\csname cnt@rev.pol.keep.sspro.p2.generic-ppi-source@0.25.keep20\endcsname{1}
\expandafter\def\csname pct@rev.pol.keep.sspro.p2.generic-ppi-source@0.25.keep20\endcsname{86.2}
\expandafter\def\csname ppt@rev.pol.keep.sspro.p2.generic-ppi-source@0.25.keep20\endcsname{$+$86.2}
\expandafter\def\csname val@rev.pol.keep.sspro.p2.generic-ppi-source@0.25.keepN\endcsname{0.898}
\expandafter\def\csname cnt@rev.pol.keep.sspro.p2.generic-ppi-source@0.25.keepN\endcsname{1}
\expandafter\def\csname pct@rev.pol.keep.sspro.p2.generic-ppi-source@0.25.keepN\endcsname{89.8}
\expandafter\def\csname ppt@rev.pol.keep.sspro.p2.generic-ppi-source@0.25.keepN\endcsname{$+$89.8}
\expandafter\def\csname val@rev.pol.keep.sspro.p2.generic-ppi-source@0.25.scoreN\endcsname{0.891}
\expandafter\def\csname cnt@rev.pol.keep.sspro.p2.generic-ppi-source@0.25.scoreN\endcsname{1}
\expandafter\def\csname pct@rev.pol.keep.sspro.p2.generic-ppi-source@0.25.scoreN\endcsname{89.1}
\expandafter\def\csname ppt@rev.pol.keep.sspro.p2.generic-ppi-source@0.25.scoreN\endcsname{$+$89.1}
\expandafter\def\csname val@rev.pol.keep.sspro.p2.generic-ppi-source@0.5.infeasible-rateN\endcsname{0.000}
\expandafter\def\csname cnt@rev.pol.keep.sspro.p2.generic-ppi-source@0.5.infeasible-rateN\endcsname{0}
\expandafter\def\csname pct@rev.pol.keep.sspro.p2.generic-ppi-source@0.5.infeasible-rateN\endcsname{0.0}
\expandafter\def\csname ppt@rev.pol.keep.sspro.p2.generic-ppi-source@0.5.infeasible-rateN\endcsname{0.0}
\expandafter\def\csname val@rev.pol.keep.sspro.p2.generic-ppi-source@0.5.keep20\endcsname{0.950}
\expandafter\def\csname cnt@rev.pol.keep.sspro.p2.generic-ppi-source@0.5.keep20\endcsname{1}
\expandafter\def\csname pct@rev.pol.keep.sspro.p2.generic-ppi-source@0.5.keep20\endcsname{95.0}
\expandafter\def\csname ppt@rev.pol.keep.sspro.p2.generic-ppi-source@0.5.keep20\endcsname{$+$95.0}
\expandafter\def\csname val@rev.pol.keep.sspro.p2.generic-ppi-source@0.5.keepN\endcsname{0.971}
\expandafter\def\csname cnt@rev.pol.keep.sspro.p2.generic-ppi-source@0.5.keepN\endcsname{1}
\expandafter\def\csname pct@rev.pol.keep.sspro.p2.generic-ppi-source@0.5.keepN\endcsname{97.1}
\expandafter\def\csname ppt@rev.pol.keep.sspro.p2.generic-ppi-source@0.5.keepN\endcsname{$+$97.1}
\expandafter\def\csname val@rev.pol.keep.sspro.p2.generic-ppi-source@0.5.scoreN\endcsname{0.902}
\expandafter\def\csname cnt@rev.pol.keep.sspro.p2.generic-ppi-source@0.5.scoreN\endcsname{1}
\expandafter\def\csname pct@rev.pol.keep.sspro.p2.generic-ppi-source@0.5.scoreN\endcsname{90.2}
\expandafter\def\csname ppt@rev.pol.keep.sspro.p2.generic-ppi-source@0.5.scoreN\endcsname{$+$90.2}
\expandafter\def\csname val@rev.pol.keep.sspro.p2.generic-ppi-source@1.0.infeasible-rateN\endcsname{0.000}
\expandafter\def\csname cnt@rev.pol.keep.sspro.p2.generic-ppi-source@1.0.infeasible-rateN\endcsname{0}
\expandafter\def\csname pct@rev.pol.keep.sspro.p2.generic-ppi-source@1.0.infeasible-rateN\endcsname{0.0}
\expandafter\def\csname ppt@rev.pol.keep.sspro.p2.generic-ppi-source@1.0.infeasible-rateN\endcsname{0.0}
\expandafter\def\csname val@rev.pol.keep.sspro.p2.generic-ppi-source@1.0.keep20\endcsname{1.000}
\expandafter\def\csname cnt@rev.pol.keep.sspro.p2.generic-ppi-source@1.0.keep20\endcsname{1}
\expandafter\def\csname pct@rev.pol.keep.sspro.p2.generic-ppi-source@1.0.keep20\endcsname{100.0}
\expandafter\def\csname ppt@rev.pol.keep.sspro.p2.generic-ppi-source@1.0.keep20\endcsname{$+$100.0}
\expandafter\def\csname val@rev.pol.keep.sspro.p2.generic-ppi-source@1.0.keepN\endcsname{1.000}
\expandafter\def\csname cnt@rev.pol.keep.sspro.p2.generic-ppi-source@1.0.keepN\endcsname{1}
\expandafter\def\csname pct@rev.pol.keep.sspro.p2.generic-ppi-source@1.0.keepN\endcsname{100.0}
\expandafter\def\csname ppt@rev.pol.keep.sspro.p2.generic-ppi-source@1.0.keepN\endcsname{$+$100.0}
\expandafter\def\csname val@rev.pol.keep.sspro.p2.generic-ppi-source@1.0.scoreN\endcsname{0.906}
\expandafter\def\csname cnt@rev.pol.keep.sspro.p2.generic-ppi-source@1.0.scoreN\endcsname{1}
\expandafter\def\csname pct@rev.pol.keep.sspro.p2.generic-ppi-source@1.0.scoreN\endcsname{90.6}
\expandafter\def\csname ppt@rev.pol.keep.sspro.p2.generic-ppi-source@1.0.scoreN\endcsname{$+$90.6}
\expandafter\def\csname val@rev.pol.keep.sspro.p2.keep-all.keep20\endcsname{1.000}
\expandafter\def\csname cnt@rev.pol.keep.sspro.p2.keep-all.keep20\endcsname{1}
\expandafter\def\csname pct@rev.pol.keep.sspro.p2.keep-all.keep20\endcsname{100.0}
\expandafter\def\csname ppt@rev.pol.keep.sspro.p2.keep-all.keep20\endcsname{$+$100.0}
\expandafter\def\csname val@rev.pol.keep.sspro.p2.keep-all.keepN\endcsname{1.000}
\expandafter\def\csname cnt@rev.pol.keep.sspro.p2.keep-all.keepN\endcsname{1}
\expandafter\def\csname pct@rev.pol.keep.sspro.p2.keep-all.keepN\endcsname{100.0}
\expandafter\def\csname ppt@rev.pol.keep.sspro.p2.keep-all.keepN\endcsname{$+$100.0}
\expandafter\def\csname val@rev.pol.keep.sspro.p2.keep-all.scoreN\endcsname{0.906}
\expandafter\def\csname cnt@rev.pol.keep.sspro.p2.keep-all.scoreN\endcsname{1}
\expandafter\def\csname pct@rev.pol.keep.sspro.p2.keep-all.scoreN\endcsname{90.6}
\expandafter\def\csname ppt@rev.pol.keep.sspro.p2.keep-all.scoreN\endcsname{$+$90.6}
\expandafter\def\csname val@rev.pol.keep.sspro.p2.old-decision.keep20\endcsname{1.000}
\expandafter\def\csname cnt@rev.pol.keep.sspro.p2.old-decision.keep20\endcsname{1}
\expandafter\def\csname pct@rev.pol.keep.sspro.p2.old-decision.keep20\endcsname{100.0}
\expandafter\def\csname ppt@rev.pol.keep.sspro.p2.old-decision.keep20\endcsname{$+$100.0}
\expandafter\def\csname val@rev.pol.keep.sspro.p2.old-decision.keepN\endcsname{1.000}
\expandafter\def\csname cnt@rev.pol.keep.sspro.p2.old-decision.keepN\endcsname{1}
\expandafter\def\csname pct@rev.pol.keep.sspro.p2.old-decision.keepN\endcsname{100.0}
\expandafter\def\csname ppt@rev.pol.keep.sspro.p2.old-decision.keepN\endcsname{$+$100.0}
\expandafter\def\csname val@rev.pol.keep.sspro.p2.old-decision.scoreN\endcsname{0.906}
\expandafter\def\csname cnt@rev.pol.keep.sspro.p2.old-decision.scoreN\endcsname{1}
\expandafter\def\csname pct@rev.pol.keep.sspro.p2.old-decision.scoreN\endcsname{90.6}
\expandafter\def\csname ppt@rev.pol.keep.sspro.p2.old-decision.scoreN\endcsname{$+$90.6}
\expandafter\def\csname val@rev.pol.keep.sspro.p2.oracle-on-T.keep20\endcsname{1.000}
\expandafter\def\csname cnt@rev.pol.keep.sspro.p2.oracle-on-T.keep20\endcsname{1}
\expandafter\def\csname pct@rev.pol.keep.sspro.p2.oracle-on-T.keep20\endcsname{100.0}
\expandafter\def\csname ppt@rev.pol.keep.sspro.p2.oracle-on-T.keep20\endcsname{$+$100.0}
\expandafter\def\csname val@rev.pol.keep.sspro.p2.oracle-on-T.keepN\endcsname{1.000}
\expandafter\def\csname cnt@rev.pol.keep.sspro.p2.oracle-on-T.keepN\endcsname{1}
\expandafter\def\csname pct@rev.pol.keep.sspro.p2.oracle-on-T.keepN\endcsname{100.0}
\expandafter\def\csname ppt@rev.pol.keep.sspro.p2.oracle-on-T.keepN\endcsname{$+$100.0}
\expandafter\def\csname val@rev.pol.keep.sspro.p2.oracle-on-T.scoreN\endcsname{0.906}
\expandafter\def\csname cnt@rev.pol.keep.sspro.p2.oracle-on-T.scoreN\endcsname{1}
\expandafter\def\csname pct@rev.pol.keep.sspro.p2.oracle-on-T.scoreN\endcsname{90.6}
\expandafter\def\csname ppt@rev.pol.keep.sspro.p2.oracle-on-T.scoreN\endcsname{$+$90.6}
\expandafter\def\csname val@rev.pol.keep.sspro.p2.rcr-stratified@0.1.infeasible-rateN\endcsname{0.250}
\expandafter\def\csname cnt@rev.pol.keep.sspro.p2.rcr-stratified@0.1.infeasible-rateN\endcsname{0}
\expandafter\def\csname pct@rev.pol.keep.sspro.p2.rcr-stratified@0.1.infeasible-rateN\endcsname{25.0}
\expandafter\def\csname ppt@rev.pol.keep.sspro.p2.rcr-stratified@0.1.infeasible-rateN\endcsname{$+$25.0}
\expandafter\def\csname val@rev.pol.keep.sspro.p2.rcr-stratified@0.1.keep20\endcsname{0.662}
\expandafter\def\csname cnt@rev.pol.keep.sspro.p2.rcr-stratified@0.1.keep20\endcsname{1}
\expandafter\def\csname pct@rev.pol.keep.sspro.p2.rcr-stratified@0.1.keep20\endcsname{66.2}
\expandafter\def\csname ppt@rev.pol.keep.sspro.p2.rcr-stratified@0.1.keep20\endcsname{$+$66.2}
\expandafter\def\csname val@rev.pol.keep.sspro.p2.rcr-stratified@0.1.keepN\endcsname{0.660}
\expandafter\def\csname cnt@rev.pol.keep.sspro.p2.rcr-stratified@0.1.keepN\endcsname{1}
\expandafter\def\csname pct@rev.pol.keep.sspro.p2.rcr-stratified@0.1.keepN\endcsname{66.0}
\expandafter\def\csname ppt@rev.pol.keep.sspro.p2.rcr-stratified@0.1.keepN\endcsname{$+$66.0}
\expandafter\def\csname val@rev.pol.keep.sspro.p2.rcr-stratified@0.1.scoreN\endcsname{0.856}
\expandafter\def\csname cnt@rev.pol.keep.sspro.p2.rcr-stratified@0.1.scoreN\endcsname{1}
\expandafter\def\csname pct@rev.pol.keep.sspro.p2.rcr-stratified@0.1.scoreN\endcsname{85.6}
\expandafter\def\csname ppt@rev.pol.keep.sspro.p2.rcr-stratified@0.1.scoreN\endcsname{$+$85.6}
\expandafter\def\csname val@rev.pol.keep.sspro.p2.rcr-stratified@0.25.infeasible-rateN\endcsname{0.000}
\expandafter\def\csname cnt@rev.pol.keep.sspro.p2.rcr-stratified@0.25.infeasible-rateN\endcsname{0}
\expandafter\def\csname pct@rev.pol.keep.sspro.p2.rcr-stratified@0.25.infeasible-rateN\endcsname{0.0}
\expandafter\def\csname ppt@rev.pol.keep.sspro.p2.rcr-stratified@0.25.infeasible-rateN\endcsname{0.0}
\expandafter\def\csname val@rev.pol.keep.sspro.p2.rcr-stratified@0.25.keep20\endcsname{0.925}
\expandafter\def\csname cnt@rev.pol.keep.sspro.p2.rcr-stratified@0.25.keep20\endcsname{1}
\expandafter\def\csname pct@rev.pol.keep.sspro.p2.rcr-stratified@0.25.keep20\endcsname{92.5}
\expandafter\def\csname ppt@rev.pol.keep.sspro.p2.rcr-stratified@0.25.keep20\endcsname{$+$92.5}
\expandafter\def\csname val@rev.pol.keep.sspro.p2.rcr-stratified@0.25.keepN\endcsname{0.897}
\expandafter\def\csname cnt@rev.pol.keep.sspro.p2.rcr-stratified@0.25.keepN\endcsname{1}
\expandafter\def\csname pct@rev.pol.keep.sspro.p2.rcr-stratified@0.25.keepN\endcsname{89.7}
\expandafter\def\csname ppt@rev.pol.keep.sspro.p2.rcr-stratified@0.25.keepN\endcsname{$+$89.7}
\expandafter\def\csname val@rev.pol.keep.sspro.p2.rcr-stratified@0.25.scoreN\endcsname{0.891}
\expandafter\def\csname cnt@rev.pol.keep.sspro.p2.rcr-stratified@0.25.scoreN\endcsname{1}
\expandafter\def\csname pct@rev.pol.keep.sspro.p2.rcr-stratified@0.25.scoreN\endcsname{89.1}
\expandafter\def\csname ppt@rev.pol.keep.sspro.p2.rcr-stratified@0.25.scoreN\endcsname{$+$89.1}
\expandafter\def\csname val@rev.pol.keep.sspro.p2.rcr-stratified@0.5.infeasible-rateN\endcsname{0.000}
\expandafter\def\csname cnt@rev.pol.keep.sspro.p2.rcr-stratified@0.5.infeasible-rateN\endcsname{0}
\expandafter\def\csname pct@rev.pol.keep.sspro.p2.rcr-stratified@0.5.infeasible-rateN\endcsname{0.0}
\expandafter\def\csname ppt@rev.pol.keep.sspro.p2.rcr-stratified@0.5.infeasible-rateN\endcsname{0.0}
\expandafter\def\csname val@rev.pol.keep.sspro.p2.rcr-stratified@0.5.keep20\endcsname{0.975}
\expandafter\def\csname cnt@rev.pol.keep.sspro.p2.rcr-stratified@0.5.keep20\endcsname{1}
\expandafter\def\csname pct@rev.pol.keep.sspro.p2.rcr-stratified@0.5.keep20\endcsname{97.5}
\expandafter\def\csname ppt@rev.pol.keep.sspro.p2.rcr-stratified@0.5.keep20\endcsname{$+$97.5}
\expandafter\def\csname val@rev.pol.keep.sspro.p2.rcr-stratified@0.5.keepN\endcsname{0.961}
\expandafter\def\csname cnt@rev.pol.keep.sspro.p2.rcr-stratified@0.5.keepN\endcsname{1}
\expandafter\def\csname pct@rev.pol.keep.sspro.p2.rcr-stratified@0.5.keepN\endcsname{96.1}
\expandafter\def\csname ppt@rev.pol.keep.sspro.p2.rcr-stratified@0.5.keepN\endcsname{$+$96.1}
\expandafter\def\csname val@rev.pol.keep.sspro.p2.rcr-stratified@0.5.scoreN\endcsname{0.900}
\expandafter\def\csname cnt@rev.pol.keep.sspro.p2.rcr-stratified@0.5.scoreN\endcsname{1}
\expandafter\def\csname pct@rev.pol.keep.sspro.p2.rcr-stratified@0.5.scoreN\endcsname{90.0}
\expandafter\def\csname ppt@rev.pol.keep.sspro.p2.rcr-stratified@0.5.scoreN\endcsname{$+$90.0}
\expandafter\def\csname val@rev.pol.keep.sspro.p2.rcr-stratified@1.0.infeasible-rateN\endcsname{0.000}
\expandafter\def\csname cnt@rev.pol.keep.sspro.p2.rcr-stratified@1.0.infeasible-rateN\endcsname{0}
\expandafter\def\csname pct@rev.pol.keep.sspro.p2.rcr-stratified@1.0.infeasible-rateN\endcsname{0.0}
\expandafter\def\csname ppt@rev.pol.keep.sspro.p2.rcr-stratified@1.0.infeasible-rateN\endcsname{0.0}
\expandafter\def\csname val@rev.pol.keep.sspro.p2.rcr-stratified@1.0.keep20\endcsname{1.000}
\expandafter\def\csname cnt@rev.pol.keep.sspro.p2.rcr-stratified@1.0.keep20\endcsname{1}
\expandafter\def\csname pct@rev.pol.keep.sspro.p2.rcr-stratified@1.0.keep20\endcsname{100.0}
\expandafter\def\csname ppt@rev.pol.keep.sspro.p2.rcr-stratified@1.0.keep20\endcsname{$+$100.0}
\expandafter\def\csname val@rev.pol.keep.sspro.p2.rcr-stratified@1.0.keepN\endcsname{1.000}
\expandafter\def\csname cnt@rev.pol.keep.sspro.p2.rcr-stratified@1.0.keepN\endcsname{1}
\expandafter\def\csname pct@rev.pol.keep.sspro.p2.rcr-stratified@1.0.keepN\endcsname{100.0}
\expandafter\def\csname ppt@rev.pol.keep.sspro.p2.rcr-stratified@1.0.keepN\endcsname{$+$100.0}
\expandafter\def\csname val@rev.pol.keep.sspro.p2.rcr-stratified@1.0.scoreN\endcsname{0.906}
\expandafter\def\csname cnt@rev.pol.keep.sspro.p2.rcr-stratified@1.0.scoreN\endcsname{1}
\expandafter\def\csname pct@rev.pol.keep.sspro.p2.rcr-stratified@1.0.scoreN\endcsname{90.6}
\expandafter\def\csname ppt@rev.pol.keep.sspro.p2.rcr-stratified@1.0.scoreN\endcsname{$+$90.6}
\expandafter\def\csname val@rev.pol.keep.sspro.p2.replay-only.keep20\endcsname{1.000}
\expandafter\def\csname cnt@rev.pol.keep.sspro.p2.replay-only.keep20\endcsname{1}
\expandafter\def\csname pct@rev.pol.keep.sspro.p2.replay-only.keep20\endcsname{100.0}
\expandafter\def\csname ppt@rev.pol.keep.sspro.p2.replay-only.keep20\endcsname{$+$100.0}
\expandafter\def\csname val@rev.pol.keep.sspro.p2.replay-only.keepN\endcsname{1.000}
\expandafter\def\csname cnt@rev.pol.keep.sspro.p2.replay-only.keepN\endcsname{1}
\expandafter\def\csname pct@rev.pol.keep.sspro.p2.replay-only.keepN\endcsname{100.0}
\expandafter\def\csname ppt@rev.pol.keep.sspro.p2.replay-only.keepN\endcsname{$+$100.0}
\expandafter\def\csname val@rev.pol.keep.sspro.p2.replay-only.scoreN\endcsname{0.906}
\expandafter\def\csname cnt@rev.pol.keep.sspro.p2.replay-only.scoreN\endcsname{1}
\expandafter\def\csname pct@rev.pol.keep.sspro.p2.replay-only.scoreN\endcsname{90.6}
\expandafter\def\csname ppt@rev.pol.keep.sspro.p2.replay-only.scoreN\endcsname{$+$90.6}
\expandafter\def\csname val@rev.pol.keep.sspro.p2.replay-ppi-uniform@0.1.infeasible-rateN\endcsname{0.000}
\expandafter\def\csname cnt@rev.pol.keep.sspro.p2.replay-ppi-uniform@0.1.infeasible-rateN\endcsname{0}
\expandafter\def\csname pct@rev.pol.keep.sspro.p2.replay-ppi-uniform@0.1.infeasible-rateN\endcsname{0.0}
\expandafter\def\csname ppt@rev.pol.keep.sspro.p2.replay-ppi-uniform@0.1.infeasible-rateN\endcsname{0.0}
\expandafter\def\csname val@rev.pol.keep.sspro.p2.replay-ppi-uniform@0.1.keep20\endcsname{0.887}
\expandafter\def\csname cnt@rev.pol.keep.sspro.p2.replay-ppi-uniform@0.1.keep20\endcsname{1}
\expandafter\def\csname pct@rev.pol.keep.sspro.p2.replay-ppi-uniform@0.1.keep20\endcsname{88.8}
\expandafter\def\csname ppt@rev.pol.keep.sspro.p2.replay-ppi-uniform@0.1.keep20\endcsname{$+$88.8}
\expandafter\def\csname val@rev.pol.keep.sspro.p2.replay-ppi-uniform@0.1.keepN\endcsname{0.891}
\expandafter\def\csname cnt@rev.pol.keep.sspro.p2.replay-ppi-uniform@0.1.keepN\endcsname{1}
\expandafter\def\csname pct@rev.pol.keep.sspro.p2.replay-ppi-uniform@0.1.keepN\endcsname{89.1}
\expandafter\def\csname ppt@rev.pol.keep.sspro.p2.replay-ppi-uniform@0.1.keepN\endcsname{$+$89.1}
\expandafter\def\csname val@rev.pol.keep.sspro.p2.replay-ppi-uniform@0.1.scoreN\endcsname{0.890}
\expandafter\def\csname cnt@rev.pol.keep.sspro.p2.replay-ppi-uniform@0.1.scoreN\endcsname{1}
\expandafter\def\csname pct@rev.pol.keep.sspro.p2.replay-ppi-uniform@0.1.scoreN\endcsname{89.0}
\expandafter\def\csname ppt@rev.pol.keep.sspro.p2.replay-ppi-uniform@0.1.scoreN\endcsname{$+$89.0}
\expandafter\def\csname val@rev.pol.keep.sspro.p2.replay-ppi-uniform@0.25.infeasible-rateN\endcsname{0.000}
\expandafter\def\csname cnt@rev.pol.keep.sspro.p2.replay-ppi-uniform@0.25.infeasible-rateN\endcsname{0}
\expandafter\def\csname pct@rev.pol.keep.sspro.p2.replay-ppi-uniform@0.25.infeasible-rateN\endcsname{0.0}
\expandafter\def\csname ppt@rev.pol.keep.sspro.p2.replay-ppi-uniform@0.25.infeasible-rateN\endcsname{0.0}
\expandafter\def\csname val@rev.pol.keep.sspro.p2.replay-ppi-uniform@0.25.keep20\endcsname{0.863}
\expandafter\def\csname cnt@rev.pol.keep.sspro.p2.replay-ppi-uniform@0.25.keep20\endcsname{1}
\expandafter\def\csname pct@rev.pol.keep.sspro.p2.replay-ppi-uniform@0.25.keep20\endcsname{86.2}
\expandafter\def\csname ppt@rev.pol.keep.sspro.p2.replay-ppi-uniform@0.25.keep20\endcsname{$+$86.2}
\expandafter\def\csname val@rev.pol.keep.sspro.p2.replay-ppi-uniform@0.25.keepN\endcsname{0.897}
\expandafter\def\csname cnt@rev.pol.keep.sspro.p2.replay-ppi-uniform@0.25.keepN\endcsname{1}
\expandafter\def\csname pct@rev.pol.keep.sspro.p2.replay-ppi-uniform@0.25.keepN\endcsname{89.7}
\expandafter\def\csname ppt@rev.pol.keep.sspro.p2.replay-ppi-uniform@0.25.keepN\endcsname{$+$89.7}
\expandafter\def\csname val@rev.pol.keep.sspro.p2.replay-ppi-uniform@0.25.scoreN\endcsname{0.891}
\expandafter\def\csname cnt@rev.pol.keep.sspro.p2.replay-ppi-uniform@0.25.scoreN\endcsname{1}
\expandafter\def\csname pct@rev.pol.keep.sspro.p2.replay-ppi-uniform@0.25.scoreN\endcsname{89.1}
\expandafter\def\csname ppt@rev.pol.keep.sspro.p2.replay-ppi-uniform@0.25.scoreN\endcsname{$+$89.1}
\expandafter\def\csname val@rev.pol.keep.sspro.p2.replay-ppi-uniform@0.5.infeasible-rateN\endcsname{0.000}
\expandafter\def\csname cnt@rev.pol.keep.sspro.p2.replay-ppi-uniform@0.5.infeasible-rateN\endcsname{0}
\expandafter\def\csname pct@rev.pol.keep.sspro.p2.replay-ppi-uniform@0.5.infeasible-rateN\endcsname{0.0}
\expandafter\def\csname ppt@rev.pol.keep.sspro.p2.replay-ppi-uniform@0.5.infeasible-rateN\endcsname{0.0}
\expandafter\def\csname val@rev.pol.keep.sspro.p2.replay-ppi-uniform@0.5.keep20\endcsname{0.938}
\expandafter\def\csname cnt@rev.pol.keep.sspro.p2.replay-ppi-uniform@0.5.keep20\endcsname{1}
\expandafter\def\csname pct@rev.pol.keep.sspro.p2.replay-ppi-uniform@0.5.keep20\endcsname{93.8}
\expandafter\def\csname ppt@rev.pol.keep.sspro.p2.replay-ppi-uniform@0.5.keep20\endcsname{$+$93.8}
\expandafter\def\csname val@rev.pol.keep.sspro.p2.replay-ppi-uniform@0.5.keepN\endcsname{0.971}
\expandafter\def\csname cnt@rev.pol.keep.sspro.p2.replay-ppi-uniform@0.5.keepN\endcsname{1}
\expandafter\def\csname pct@rev.pol.keep.sspro.p2.replay-ppi-uniform@0.5.keepN\endcsname{97.1}
\expandafter\def\csname ppt@rev.pol.keep.sspro.p2.replay-ppi-uniform@0.5.keepN\endcsname{$+$97.1}
\expandafter\def\csname val@rev.pol.keep.sspro.p2.replay-ppi-uniform@0.5.scoreN\endcsname{0.902}
\expandafter\def\csname cnt@rev.pol.keep.sspro.p2.replay-ppi-uniform@0.5.scoreN\endcsname{1}
\expandafter\def\csname pct@rev.pol.keep.sspro.p2.replay-ppi-uniform@0.5.scoreN\endcsname{90.2}
\expandafter\def\csname ppt@rev.pol.keep.sspro.p2.replay-ppi-uniform@0.5.scoreN\endcsname{$+$90.2}
\expandafter\def\csname val@rev.pol.keep.sspro.p2.replay-ppi-uniform@1.0.infeasible-rateN\endcsname{0.000}
\expandafter\def\csname cnt@rev.pol.keep.sspro.p2.replay-ppi-uniform@1.0.infeasible-rateN\endcsname{0}
\expandafter\def\csname pct@rev.pol.keep.sspro.p2.replay-ppi-uniform@1.0.infeasible-rateN\endcsname{0.0}
\expandafter\def\csname ppt@rev.pol.keep.sspro.p2.replay-ppi-uniform@1.0.infeasible-rateN\endcsname{0.0}
\expandafter\def\csname val@rev.pol.keep.sspro.p2.replay-ppi-uniform@1.0.keep20\endcsname{1.000}
\expandafter\def\csname cnt@rev.pol.keep.sspro.p2.replay-ppi-uniform@1.0.keep20\endcsname{1}
\expandafter\def\csname pct@rev.pol.keep.sspro.p2.replay-ppi-uniform@1.0.keep20\endcsname{100.0}
\expandafter\def\csname ppt@rev.pol.keep.sspro.p2.replay-ppi-uniform@1.0.keep20\endcsname{$+$100.0}
\expandafter\def\csname val@rev.pol.keep.sspro.p2.replay-ppi-uniform@1.0.keepN\endcsname{1.000}
\expandafter\def\csname cnt@rev.pol.keep.sspro.p2.replay-ppi-uniform@1.0.keepN\endcsname{1}
\expandafter\def\csname pct@rev.pol.keep.sspro.p2.replay-ppi-uniform@1.0.keepN\endcsname{100.0}
\expandafter\def\csname ppt@rev.pol.keep.sspro.p2.replay-ppi-uniform@1.0.keepN\endcsname{$+$100.0}
\expandafter\def\csname val@rev.pol.keep.sspro.p2.replay-ppi-uniform@1.0.scoreN\endcsname{0.906}
\expandafter\def\csname cnt@rev.pol.keep.sspro.p2.replay-ppi-uniform@1.0.scoreN\endcsname{1}
\expandafter\def\csname pct@rev.pol.keep.sspro.p2.replay-ppi-uniform@1.0.scoreN\endcsname{90.6}
\expandafter\def\csname ppt@rev.pol.keep.sspro.p2.replay-ppi-uniform@1.0.scoreN\endcsname{$+$90.6}
\expandafter\def\csname val@rev.pol.keep.sspro.p2.suspend-all.keep20\endcsname{0.000}
\expandafter\def\csname cnt@rev.pol.keep.sspro.p2.suspend-all.keep20\endcsname{0}
\expandafter\def\csname pct@rev.pol.keep.sspro.p2.suspend-all.keep20\endcsname{0.0}
\expandafter\def\csname ppt@rev.pol.keep.sspro.p2.suspend-all.keep20\endcsname{0.0}
\expandafter\def\csname val@rev.pol.keep.sspro.p2.suspend-all.keepN\endcsname{0.000}
\expandafter\def\csname cnt@rev.pol.keep.sspro.p2.suspend-all.keepN\endcsname{0}
\expandafter\def\csname pct@rev.pol.keep.sspro.p2.suspend-all.keepN\endcsname{0.0}
\expandafter\def\csname ppt@rev.pol.keep.sspro.p2.suspend-all.keepN\endcsname{0.0}
\expandafter\def\csname val@rev.pol.keep.sspro.p2.suspend-all.scoreN\endcsname{0.760}
\expandafter\def\csname cnt@rev.pol.keep.sspro.p2.suspend-all.scoreN\endcsname{1}
\expandafter\def\csname pct@rev.pol.keep.sspro.p2.suspend-all.scoreN\endcsname{76.0}
\expandafter\def\csname ppt@rev.pol.keep.sspro.p2.suspend-all.scoreN\endcsname{$+$76.0}
\expandafter\def\csname val@rev.pol.keep.sspro.p2.uniform-fresh@0.1.infeasible-rateN\endcsname{0.000}
\expandafter\def\csname cnt@rev.pol.keep.sspro.p2.uniform-fresh@0.1.infeasible-rateN\endcsname{0}
\expandafter\def\csname pct@rev.pol.keep.sspro.p2.uniform-fresh@0.1.infeasible-rateN\endcsname{0.0}
\expandafter\def\csname ppt@rev.pol.keep.sspro.p2.uniform-fresh@0.1.infeasible-rateN\endcsname{0.0}
\expandafter\def\csname val@rev.pol.keep.sspro.p2.uniform-fresh@0.1.keep20\endcsname{0.925}
\expandafter\def\csname cnt@rev.pol.keep.sspro.p2.uniform-fresh@0.1.keep20\endcsname{1}
\expandafter\def\csname pct@rev.pol.keep.sspro.p2.uniform-fresh@0.1.keep20\endcsname{92.5}
\expandafter\def\csname ppt@rev.pol.keep.sspro.p2.uniform-fresh@0.1.keep20\endcsname{$+$92.5}
\expandafter\def\csname val@rev.pol.keep.sspro.p2.uniform-fresh@0.1.keepN\endcsname{0.916}
\expandafter\def\csname cnt@rev.pol.keep.sspro.p2.uniform-fresh@0.1.keepN\endcsname{1}
\expandafter\def\csname pct@rev.pol.keep.sspro.p2.uniform-fresh@0.1.keepN\endcsname{91.6}
\expandafter\def\csname ppt@rev.pol.keep.sspro.p2.uniform-fresh@0.1.keepN\endcsname{$+$91.6}
\expandafter\def\csname val@rev.pol.keep.sspro.p2.uniform-fresh@0.1.scoreN\endcsname{0.894}
\expandafter\def\csname cnt@rev.pol.keep.sspro.p2.uniform-fresh@0.1.scoreN\endcsname{1}
\expandafter\def\csname pct@rev.pol.keep.sspro.p2.uniform-fresh@0.1.scoreN\endcsname{89.4}
\expandafter\def\csname ppt@rev.pol.keep.sspro.p2.uniform-fresh@0.1.scoreN\endcsname{$+$89.4}
\expandafter\def\csname val@rev.pol.keep.sspro.p2.uniform-fresh@0.25.infeasible-rateN\endcsname{0.000}
\expandafter\def\csname cnt@rev.pol.keep.sspro.p2.uniform-fresh@0.25.infeasible-rateN\endcsname{0}
\expandafter\def\csname pct@rev.pol.keep.sspro.p2.uniform-fresh@0.25.infeasible-rateN\endcsname{0.0}
\expandafter\def\csname ppt@rev.pol.keep.sspro.p2.uniform-fresh@0.25.infeasible-rateN\endcsname{0.0}
\expandafter\def\csname val@rev.pol.keep.sspro.p2.uniform-fresh@0.25.keep20\endcsname{0.900}
\expandafter\def\csname cnt@rev.pol.keep.sspro.p2.uniform-fresh@0.25.keep20\endcsname{1}
\expandafter\def\csname pct@rev.pol.keep.sspro.p2.uniform-fresh@0.25.keep20\endcsname{90.0}
\expandafter\def\csname ppt@rev.pol.keep.sspro.p2.uniform-fresh@0.25.keep20\endcsname{$+$90.0}
\expandafter\def\csname val@rev.pol.keep.sspro.p2.uniform-fresh@0.25.keepN\endcsname{0.925}
\expandafter\def\csname cnt@rev.pol.keep.sspro.p2.uniform-fresh@0.25.keepN\endcsname{1}
\expandafter\def\csname pct@rev.pol.keep.sspro.p2.uniform-fresh@0.25.keepN\endcsname{92.5}
\expandafter\def\csname ppt@rev.pol.keep.sspro.p2.uniform-fresh@0.25.keepN\endcsname{$+$92.5}
\expandafter\def\csname val@rev.pol.keep.sspro.p2.uniform-fresh@0.25.scoreN\endcsname{0.895}
\expandafter\def\csname cnt@rev.pol.keep.sspro.p2.uniform-fresh@0.25.scoreN\endcsname{1}
\expandafter\def\csname pct@rev.pol.keep.sspro.p2.uniform-fresh@0.25.scoreN\endcsname{89.5}
\expandafter\def\csname ppt@rev.pol.keep.sspro.p2.uniform-fresh@0.25.scoreN\endcsname{$+$89.5}
\expandafter\def\csname val@rev.pol.keep.sspro.p2.uniform-fresh@0.5.infeasible-rateN\endcsname{0.000}
\expandafter\def\csname cnt@rev.pol.keep.sspro.p2.uniform-fresh@0.5.infeasible-rateN\endcsname{0}
\expandafter\def\csname pct@rev.pol.keep.sspro.p2.uniform-fresh@0.5.infeasible-rateN\endcsname{0.0}
\expandafter\def\csname ppt@rev.pol.keep.sspro.p2.uniform-fresh@0.5.infeasible-rateN\endcsname{0.0}
\expandafter\def\csname val@rev.pol.keep.sspro.p2.uniform-fresh@0.5.keep20\endcsname{0.975}
\expandafter\def\csname cnt@rev.pol.keep.sspro.p2.uniform-fresh@0.5.keep20\endcsname{1}
\expandafter\def\csname pct@rev.pol.keep.sspro.p2.uniform-fresh@0.5.keep20\endcsname{97.5}
\expandafter\def\csname ppt@rev.pol.keep.sspro.p2.uniform-fresh@0.5.keep20\endcsname{$+$97.5}
\expandafter\def\csname val@rev.pol.keep.sspro.p2.uniform-fresh@0.5.keepN\endcsname{0.963}
\expandafter\def\csname cnt@rev.pol.keep.sspro.p2.uniform-fresh@0.5.keepN\endcsname{1}
\expandafter\def\csname pct@rev.pol.keep.sspro.p2.uniform-fresh@0.5.keepN\endcsname{96.3}
\expandafter\def\csname ppt@rev.pol.keep.sspro.p2.uniform-fresh@0.5.keepN\endcsname{$+$96.3}
\expandafter\def\csname val@rev.pol.keep.sspro.p2.uniform-fresh@0.5.scoreN\endcsname{0.901}
\expandafter\def\csname cnt@rev.pol.keep.sspro.p2.uniform-fresh@0.5.scoreN\endcsname{1}
\expandafter\def\csname pct@rev.pol.keep.sspro.p2.uniform-fresh@0.5.scoreN\endcsname{90.1}
\expandafter\def\csname ppt@rev.pol.keep.sspro.p2.uniform-fresh@0.5.scoreN\endcsname{$+$90.1}
\expandafter\def\csname val@rev.pol.keep.sspro.p2.uniform-fresh@1.0.infeasible-rateN\endcsname{0.000}
\expandafter\def\csname cnt@rev.pol.keep.sspro.p2.uniform-fresh@1.0.infeasible-rateN\endcsname{0}
\expandafter\def\csname pct@rev.pol.keep.sspro.p2.uniform-fresh@1.0.infeasible-rateN\endcsname{0.0}
\expandafter\def\csname ppt@rev.pol.keep.sspro.p2.uniform-fresh@1.0.infeasible-rateN\endcsname{0.0}
\expandafter\def\csname val@rev.pol.keep.sspro.p2.uniform-fresh@1.0.keep20\endcsname{1.000}
\expandafter\def\csname cnt@rev.pol.keep.sspro.p2.uniform-fresh@1.0.keep20\endcsname{1}
\expandafter\def\csname pct@rev.pol.keep.sspro.p2.uniform-fresh@1.0.keep20\endcsname{100.0}
\expandafter\def\csname ppt@rev.pol.keep.sspro.p2.uniform-fresh@1.0.keep20\endcsname{$+$100.0}
\expandafter\def\csname val@rev.pol.keep.sspro.p2.uniform-fresh@1.0.keepN\endcsname{1.000}
\expandafter\def\csname cnt@rev.pol.keep.sspro.p2.uniform-fresh@1.0.keepN\endcsname{1}
\expandafter\def\csname pct@rev.pol.keep.sspro.p2.uniform-fresh@1.0.keepN\endcsname{100.0}
\expandafter\def\csname ppt@rev.pol.keep.sspro.p2.uniform-fresh@1.0.keepN\endcsname{$+$100.0}
\expandafter\def\csname val@rev.pol.keep.sspro.p2.uniform-fresh@1.0.scoreN\endcsname{0.906}
\expandafter\def\csname cnt@rev.pol.keep.sspro.p2.uniform-fresh@1.0.scoreN\endcsname{1}
\expandafter\def\csname pct@rev.pol.keep.sspro.p2.uniform-fresh@1.0.scoreN\endcsname{90.6}
\expandafter\def\csname ppt@rev.pol.keep.sspro.p2.uniform-fresh@1.0.scoreN\endcsname{$+$90.6}
\expandafter\def\csname val@rev.pol.keep.uiv.p1.best-single@0.1.share-direct-N\endcsname{0.286}
\expandafter\def\csname cnt@rev.pol.keep.uiv.p1.best-single@0.1.share-direct-N\endcsname{0}
\expandafter\def\csname pct@rev.pol.keep.uiv.p1.best-single@0.1.share-direct-N\endcsname{28.6}
\expandafter\def\csname ppt@rev.pol.keep.uiv.p1.best-single@0.1.share-direct-N\endcsname{$+$28.6}
\expandafter\def\csname val@rev.pol.keep.uiv.p1.best-single@0.25.share-direct-N\endcsname{0.376}
\expandafter\def\csname cnt@rev.pol.keep.uiv.p1.best-single@0.25.share-direct-N\endcsname{0}
\expandafter\def\csname pct@rev.pol.keep.uiv.p1.best-single@0.25.share-direct-N\endcsname{37.6}
\expandafter\def\csname ppt@rev.pol.keep.uiv.p1.best-single@0.25.share-direct-N\endcsname{$+$37.6}
\expandafter\def\csname val@rev.pol.keep.uiv.p1.best-single@0.5.share-direct-N\endcsname{0.521}
\expandafter\def\csname cnt@rev.pol.keep.uiv.p1.best-single@0.5.share-direct-N\endcsname{1}
\expandafter\def\csname pct@rev.pol.keep.uiv.p1.best-single@0.5.share-direct-N\endcsname{52.1}
\expandafter\def\csname ppt@rev.pol.keep.uiv.p1.best-single@0.5.share-direct-N\endcsname{$+$52.1}
\expandafter\def\csname val@rev.pol.keep.uiv.p1.best-single@1.0.share-direct-N\endcsname{1.000}
\expandafter\def\csname cnt@rev.pol.keep.uiv.p1.best-single@1.0.share-direct-N\endcsname{1}
\expandafter\def\csname pct@rev.pol.keep.uiv.p1.best-single@1.0.share-direct-N\endcsname{100.0}
\expandafter\def\csname ppt@rev.pol.keep.uiv.p1.best-single@1.0.share-direct-N\endcsname{$+$100.0}
\expandafter\def\csname val@rev.pol.keep.uiv.p1.corr.budgeted-points\endcsname{16}
\expandafter\def\csname cnt@rev.pol.keep.uiv.p1.corr.budgeted-points\endcsname{16}
\expandafter\def\csname pct@rev.pol.keep.uiv.p1.corr.budgeted-points\endcsname{1600.0}
\expandafter\def\csname ppt@rev.pol.keep.uiv.p1.corr.budgeted-points\endcsname{$+$1600.0}
\expandafter\def\csname val@rev.pol.keep.uiv.p1.corr.pearson\endcsname{-0.955}
\expandafter\def\csname cnt@rev.pol.keep.uiv.p1.corr.pearson\endcsname{-1}
\expandafter\def\csname pct@rev.pol.keep.uiv.p1.corr.pearson\endcsname{$-$95.5}
\expandafter\def\csname ppt@rev.pol.keep.uiv.p1.corr.pearson\endcsname{$-$95.5}
\expandafter\def\csname val@rev.pol.keep.uiv.p1.corr.pearson-20\endcsname{-0.968}
\expandafter\def\csname cnt@rev.pol.keep.uiv.p1.corr.pearson-20\endcsname{-1}
\expandafter\def\csname pct@rev.pol.keep.uiv.p1.corr.pearson-20\endcsname{$-$96.8}
\expandafter\def\csname ppt@rev.pol.keep.uiv.p1.corr.pearson-20\endcsname{$-$96.8}
\expandafter\def\csname val@rev.pol.keep.uiv.p1.corr.pearson-all\endcsname{-0.995}
\expandafter\def\csname cnt@rev.pol.keep.uiv.p1.corr.pearson-all\endcsname{-1}
\expandafter\def\csname pct@rev.pol.keep.uiv.p1.corr.pearson-all\endcsname{$-$99.5}
\expandafter\def\csname ppt@rev.pol.keep.uiv.p1.corr.pearson-all\endcsname{$-$99.5}
\expandafter\def\csname val@rev.pol.keep.uiv.p1.corr.slope-pt-per-keep\endcsname{-0.072}
\expandafter\def\csname cnt@rev.pol.keep.uiv.p1.corr.slope-pt-per-keep\endcsname{0}
\expandafter\def\csname pct@rev.pol.keep.uiv.p1.corr.slope-pt-per-keep\endcsname{$-$7.2}
\expandafter\def\csname ppt@rev.pol.keep.uiv.p1.corr.slope-pt-per-keep\endcsname{$-$7.2}
\expandafter\def\csname val@rev.pol.keep.uiv.p1.corr.spearman\endcsname{-0.904}
\expandafter\def\csname cnt@rev.pol.keep.uiv.p1.corr.spearman\endcsname{-1}
\expandafter\def\csname pct@rev.pol.keep.uiv.p1.corr.spearman\endcsname{$-$90.4}
\expandafter\def\csname ppt@rev.pol.keep.uiv.p1.corr.spearman\endcsname{$-$90.4}
\expandafter\def\csname val@rev.pol.keep.uiv.p1.corr.spearman-20\endcsname{-0.956}
\expandafter\def\csname cnt@rev.pol.keep.uiv.p1.corr.spearman-20\endcsname{-1}
\expandafter\def\csname pct@rev.pol.keep.uiv.p1.corr.spearman-20\endcsname{$-$95.6}
\expandafter\def\csname ppt@rev.pol.keep.uiv.p1.corr.spearman-20\endcsname{$-$95.6}
\expandafter\def\csname val@rev.pol.keep.uiv.p1.corr.spearman-all\endcsname{-0.942}
\expandafter\def\csname cnt@rev.pol.keep.uiv.p1.corr.spearman-all\endcsname{-1}
\expandafter\def\csname pct@rev.pol.keep.uiv.p1.corr.spearman-all\endcsname{$-$94.2}
\expandafter\def\csname ppt@rev.pol.keep.uiv.p1.corr.spearman-all\endcsname{$-$94.2}
\expandafter\def\csname val@rev.pol.keep.uiv.p1.corr.spearman-primary\endcsname{-1.000}
\expandafter\def\csname cnt@rev.pol.keep.uiv.p1.corr.spearman-primary\endcsname{-1}
\expandafter\def\csname pct@rev.pol.keep.uiv.p1.corr.spearman-primary\endcsname{$-$100.0}
\expandafter\def\csname ppt@rev.pol.keep.uiv.p1.corr.spearman-primary\endcsname{$-$100.0}
\expandafter\def\csname val@rev.pol.keep.uiv.p1.corr.with-nobudget-points\endcsname{21}
\expandafter\def\csname cnt@rev.pol.keep.uiv.p1.corr.with-nobudget-points\endcsname{21}
\expandafter\def\csname pct@rev.pol.keep.uiv.p1.corr.with-nobudget-points\endcsname{2100.0}
\expandafter\def\csname ppt@rev.pol.keep.uiv.p1.corr.with-nobudget-points\endcsname{$+$2100.0}
\expandafter\def\csname val@rev.pol.keep.uiv.p1.full-recert.keep20\endcsname{0.500}
\expandafter\def\csname cnt@rev.pol.keep.uiv.p1.full-recert.keep20\endcsname{0}
\expandafter\def\csname pct@rev.pol.keep.uiv.p1.full-recert.keep20\endcsname{50.0}
\expandafter\def\csname ppt@rev.pol.keep.uiv.p1.full-recert.keep20\endcsname{$+$50.0}
\expandafter\def\csname val@rev.pol.keep.uiv.p1.full-recert.keepN\endcsname{0.500}
\expandafter\def\csname cnt@rev.pol.keep.uiv.p1.full-recert.keepN\endcsname{0}
\expandafter\def\csname pct@rev.pol.keep.uiv.p1.full-recert.keepN\endcsname{50.0}
\expandafter\def\csname ppt@rev.pol.keep.uiv.p1.full-recert.keepN\endcsname{$+$50.0}
\expandafter\def\csname val@rev.pol.keep.uiv.p1.full-recert.scoreN\endcsname{0.793}
\expandafter\def\csname cnt@rev.pol.keep.uiv.p1.full-recert.scoreN\endcsname{1}
\expandafter\def\csname pct@rev.pol.keep.uiv.p1.full-recert.scoreN\endcsname{79.3}
\expandafter\def\csname ppt@rev.pol.keep.uiv.p1.full-recert.scoreN\endcsname{$+$79.3}
\expandafter\def\csname val@rev.pol.keep.uiv.p1.generic-ppi-source@0.1.infeasible-rateN\endcsname{0.000}
\expandafter\def\csname cnt@rev.pol.keep.uiv.p1.generic-ppi-source@0.1.infeasible-rateN\endcsname{0}
\expandafter\def\csname pct@rev.pol.keep.uiv.p1.generic-ppi-source@0.1.infeasible-rateN\endcsname{0.0}
\expandafter\def\csname ppt@rev.pol.keep.uiv.p1.generic-ppi-source@0.1.infeasible-rateN\endcsname{0.0}
\expandafter\def\csname val@rev.pol.keep.uiv.p1.generic-ppi-source@0.1.keep20\endcsname{0.550}
\expandafter\def\csname cnt@rev.pol.keep.uiv.p1.generic-ppi-source@0.1.keep20\endcsname{1}
\expandafter\def\csname pct@rev.pol.keep.uiv.p1.generic-ppi-source@0.1.keep20\endcsname{55.0}
\expandafter\def\csname ppt@rev.pol.keep.uiv.p1.generic-ppi-source@0.1.keep20\endcsname{$+$55.0}
\expandafter\def\csname val@rev.pol.keep.uiv.p1.generic-ppi-source@0.1.keepN\endcsname{0.514}
\expandafter\def\csname cnt@rev.pol.keep.uiv.p1.generic-ppi-source@0.1.keepN\endcsname{1}
\expandafter\def\csname pct@rev.pol.keep.uiv.p1.generic-ppi-source@0.1.keepN\endcsname{51.4}
\expandafter\def\csname ppt@rev.pol.keep.uiv.p1.generic-ppi-source@0.1.keepN\endcsname{$+$51.4}
\expandafter\def\csname val@rev.pol.keep.uiv.p1.generic-ppi-source@0.1.scoreN\endcsname{0.798}
\expandafter\def\csname cnt@rev.pol.keep.uiv.p1.generic-ppi-source@0.1.scoreN\endcsname{1}
\expandafter\def\csname pct@rev.pol.keep.uiv.p1.generic-ppi-source@0.1.scoreN\endcsname{79.8}
\expandafter\def\csname ppt@rev.pol.keep.uiv.p1.generic-ppi-source@0.1.scoreN\endcsname{$+$79.8}
\expandafter\def\csname val@rev.pol.keep.uiv.p1.generic-ppi-source@0.25.infeasible-rateN\endcsname{0.000}
\expandafter\def\csname cnt@rev.pol.keep.uiv.p1.generic-ppi-source@0.25.infeasible-rateN\endcsname{0}
\expandafter\def\csname pct@rev.pol.keep.uiv.p1.generic-ppi-source@0.25.infeasible-rateN\endcsname{0.0}
\expandafter\def\csname ppt@rev.pol.keep.uiv.p1.generic-ppi-source@0.25.infeasible-rateN\endcsname{0.0}
\expandafter\def\csname val@rev.pol.keep.uiv.p1.generic-ppi-source@0.25.keep20\endcsname{0.575}
\expandafter\def\csname cnt@rev.pol.keep.uiv.p1.generic-ppi-source@0.25.keep20\endcsname{1}
\expandafter\def\csname pct@rev.pol.keep.uiv.p1.generic-ppi-source@0.25.keep20\endcsname{57.5}
\expandafter\def\csname ppt@rev.pol.keep.uiv.p1.generic-ppi-source@0.25.keep20\endcsname{$+$57.5}
\expandafter\def\csname val@rev.pol.keep.uiv.p1.generic-ppi-source@0.25.keepN\endcsname{0.496}
\expandafter\def\csname cnt@rev.pol.keep.uiv.p1.generic-ppi-source@0.25.keepN\endcsname{0}
\expandafter\def\csname pct@rev.pol.keep.uiv.p1.generic-ppi-source@0.25.keepN\endcsname{49.6}
\expandafter\def\csname ppt@rev.pol.keep.uiv.p1.generic-ppi-source@0.25.keepN\endcsname{$+$49.6}
\expandafter\def\csname val@rev.pol.keep.uiv.p1.generic-ppi-source@0.25.scoreN\endcsname{0.797}
\expandafter\def\csname cnt@rev.pol.keep.uiv.p1.generic-ppi-source@0.25.scoreN\endcsname{1}
\expandafter\def\csname pct@rev.pol.keep.uiv.p1.generic-ppi-source@0.25.scoreN\endcsname{79.7}
\expandafter\def\csname ppt@rev.pol.keep.uiv.p1.generic-ppi-source@0.25.scoreN\endcsname{$+$79.7}
\expandafter\def\csname val@rev.pol.keep.uiv.p1.generic-ppi-source@0.5.infeasible-rateN\endcsname{0.000}
\expandafter\def\csname cnt@rev.pol.keep.uiv.p1.generic-ppi-source@0.5.infeasible-rateN\endcsname{0}
\expandafter\def\csname pct@rev.pol.keep.uiv.p1.generic-ppi-source@0.5.infeasible-rateN\endcsname{0.0}
\expandafter\def\csname ppt@rev.pol.keep.uiv.p1.generic-ppi-source@0.5.infeasible-rateN\endcsname{0.0}
\expandafter\def\csname val@rev.pol.keep.uiv.p1.generic-ppi-source@0.5.keep20\endcsname{0.438}
\expandafter\def\csname cnt@rev.pol.keep.uiv.p1.generic-ppi-source@0.5.keep20\endcsname{0}
\expandafter\def\csname pct@rev.pol.keep.uiv.p1.generic-ppi-source@0.5.keep20\endcsname{43.8}
\expandafter\def\csname ppt@rev.pol.keep.uiv.p1.generic-ppi-source@0.5.keep20\endcsname{$+$43.8}
\expandafter\def\csname val@rev.pol.keep.uiv.p1.generic-ppi-source@0.5.keepN\endcsname{0.455}
\expandafter\def\csname cnt@rev.pol.keep.uiv.p1.generic-ppi-source@0.5.keepN\endcsname{0}
\expandafter\def\csname pct@rev.pol.keep.uiv.p1.generic-ppi-source@0.5.keepN\endcsname{45.5}
\expandafter\def\csname ppt@rev.pol.keep.uiv.p1.generic-ppi-source@0.5.keepN\endcsname{$+$45.5}
\expandafter\def\csname val@rev.pol.keep.uiv.p1.generic-ppi-source@0.5.scoreN\endcsname{0.799}
\expandafter\def\csname cnt@rev.pol.keep.uiv.p1.generic-ppi-source@0.5.scoreN\endcsname{1}
\expandafter\def\csname pct@rev.pol.keep.uiv.p1.generic-ppi-source@0.5.scoreN\endcsname{79.9}
\expandafter\def\csname ppt@rev.pol.keep.uiv.p1.generic-ppi-source@0.5.scoreN\endcsname{$+$79.9}
\expandafter\def\csname val@rev.pol.keep.uiv.p1.generic-ppi-source@1.0.infeasible-rateN\endcsname{0.000}
\expandafter\def\csname cnt@rev.pol.keep.uiv.p1.generic-ppi-source@1.0.infeasible-rateN\endcsname{0}
\expandafter\def\csname pct@rev.pol.keep.uiv.p1.generic-ppi-source@1.0.infeasible-rateN\endcsname{0.0}
\expandafter\def\csname ppt@rev.pol.keep.uiv.p1.generic-ppi-source@1.0.infeasible-rateN\endcsname{0.0}
\expandafter\def\csname val@rev.pol.keep.uiv.p1.generic-ppi-source@1.0.keep20\endcsname{0.500}
\expandafter\def\csname cnt@rev.pol.keep.uiv.p1.generic-ppi-source@1.0.keep20\endcsname{0}
\expandafter\def\csname pct@rev.pol.keep.uiv.p1.generic-ppi-source@1.0.keep20\endcsname{50.0}
\expandafter\def\csname ppt@rev.pol.keep.uiv.p1.generic-ppi-source@1.0.keep20\endcsname{$+$50.0}
\expandafter\def\csname val@rev.pol.keep.uiv.p1.generic-ppi-source@1.0.keepN\endcsname{0.500}
\expandafter\def\csname cnt@rev.pol.keep.uiv.p1.generic-ppi-source@1.0.keepN\endcsname{0}
\expandafter\def\csname pct@rev.pol.keep.uiv.p1.generic-ppi-source@1.0.keepN\endcsname{50.0}
\expandafter\def\csname ppt@rev.pol.keep.uiv.p1.generic-ppi-source@1.0.keepN\endcsname{$+$50.0}
\expandafter\def\csname val@rev.pol.keep.uiv.p1.generic-ppi-source@1.0.scoreN\endcsname{0.793}
\expandafter\def\csname cnt@rev.pol.keep.uiv.p1.generic-ppi-source@1.0.scoreN\endcsname{1}
\expandafter\def\csname pct@rev.pol.keep.uiv.p1.generic-ppi-source@1.0.scoreN\endcsname{79.3}
\expandafter\def\csname ppt@rev.pol.keep.uiv.p1.generic-ppi-source@1.0.scoreN\endcsname{$+$79.3}
\expandafter\def\csname val@rev.pol.keep.uiv.p1.keep-all.keep20\endcsname{1.000}
\expandafter\def\csname cnt@rev.pol.keep.uiv.p1.keep-all.keep20\endcsname{1}
\expandafter\def\csname pct@rev.pol.keep.uiv.p1.keep-all.keep20\endcsname{100.0}
\expandafter\def\csname ppt@rev.pol.keep.uiv.p1.keep-all.keep20\endcsname{$+$100.0}
\expandafter\def\csname val@rev.pol.keep.uiv.p1.keep-all.keepN\endcsname{1.000}
\expandafter\def\csname cnt@rev.pol.keep.uiv.p1.keep-all.keepN\endcsname{1}
\expandafter\def\csname pct@rev.pol.keep.uiv.p1.keep-all.keepN\endcsname{100.0}
\expandafter\def\csname ppt@rev.pol.keep.uiv.p1.keep-all.keepN\endcsname{$+$100.0}
\expandafter\def\csname val@rev.pol.keep.uiv.p1.keep-all.scoreN\endcsname{0.760}
\expandafter\def\csname cnt@rev.pol.keep.uiv.p1.keep-all.scoreN\endcsname{1}
\expandafter\def\csname pct@rev.pol.keep.uiv.p1.keep-all.scoreN\endcsname{76.0}
\expandafter\def\csname ppt@rev.pol.keep.uiv.p1.keep-all.scoreN\endcsname{$+$76.0}
\expandafter\def\csname val@rev.pol.keep.uiv.p1.old-decision.keep20\endcsname{0.000}
\expandafter\def\csname cnt@rev.pol.keep.uiv.p1.old-decision.keep20\endcsname{0}
\expandafter\def\csname pct@rev.pol.keep.uiv.p1.old-decision.keep20\endcsname{0.0}
\expandafter\def\csname ppt@rev.pol.keep.uiv.p1.old-decision.keep20\endcsname{0.0}
\expandafter\def\csname val@rev.pol.keep.uiv.p1.old-decision.keepN\endcsname{0.000}
\expandafter\def\csname cnt@rev.pol.keep.uiv.p1.old-decision.keepN\endcsname{0}
\expandafter\def\csname pct@rev.pol.keep.uiv.p1.old-decision.keepN\endcsname{0.0}
\expandafter\def\csname ppt@rev.pol.keep.uiv.p1.old-decision.keepN\endcsname{0.0}
\expandafter\def\csname val@rev.pol.keep.uiv.p1.old-decision.scoreN\endcsname{0.833}
\expandafter\def\csname cnt@rev.pol.keep.uiv.p1.old-decision.scoreN\endcsname{1}
\expandafter\def\csname pct@rev.pol.keep.uiv.p1.old-decision.scoreN\endcsname{83.3}
\expandafter\def\csname ppt@rev.pol.keep.uiv.p1.old-decision.scoreN\endcsname{$+$83.3}
\expandafter\def\csname val@rev.pol.keep.uiv.p1.oracle-on-T.keep20\endcsname{0.000}
\expandafter\def\csname cnt@rev.pol.keep.uiv.p1.oracle-on-T.keep20\endcsname{0}
\expandafter\def\csname pct@rev.pol.keep.uiv.p1.oracle-on-T.keep20\endcsname{0.0}
\expandafter\def\csname ppt@rev.pol.keep.uiv.p1.oracle-on-T.keep20\endcsname{0.0}
\expandafter\def\csname val@rev.pol.keep.uiv.p1.oracle-on-T.keepN\endcsname{0.000}
\expandafter\def\csname cnt@rev.pol.keep.uiv.p1.oracle-on-T.keepN\endcsname{0}
\expandafter\def\csname pct@rev.pol.keep.uiv.p1.oracle-on-T.keepN\endcsname{0.0}
\expandafter\def\csname ppt@rev.pol.keep.uiv.p1.oracle-on-T.keepN\endcsname{0.0}
\expandafter\def\csname val@rev.pol.keep.uiv.p1.oracle-on-T.scoreN\endcsname{0.833}
\expandafter\def\csname cnt@rev.pol.keep.uiv.p1.oracle-on-T.scoreN\endcsname{1}
\expandafter\def\csname pct@rev.pol.keep.uiv.p1.oracle-on-T.scoreN\endcsname{83.3}
\expandafter\def\csname ppt@rev.pol.keep.uiv.p1.oracle-on-T.scoreN\endcsname{$+$83.3}
\expandafter\def\csname val@rev.pol.keep.uiv.p1.rcr-stratified@0.1.infeasible-rateN\endcsname{0.000}
\expandafter\def\csname cnt@rev.pol.keep.uiv.p1.rcr-stratified@0.1.infeasible-rateN\endcsname{0}
\expandafter\def\csname pct@rev.pol.keep.uiv.p1.rcr-stratified@0.1.infeasible-rateN\endcsname{0.0}
\expandafter\def\csname ppt@rev.pol.keep.uiv.p1.rcr-stratified@0.1.infeasible-rateN\endcsname{0.0}
\expandafter\def\csname val@rev.pol.keep.uiv.p1.rcr-stratified@0.1.keep20\endcsname{0.338}
\expandafter\def\csname cnt@rev.pol.keep.uiv.p1.rcr-stratified@0.1.keep20\endcsname{0}
\expandafter\def\csname pct@rev.pol.keep.uiv.p1.rcr-stratified@0.1.keep20\endcsname{33.8}
\expandafter\def\csname ppt@rev.pol.keep.uiv.p1.rcr-stratified@0.1.keep20\endcsname{$+$33.8}
\expandafter\def\csname val@rev.pol.keep.uiv.p1.rcr-stratified@0.1.keepN\endcsname{0.383}
\expandafter\def\csname cnt@rev.pol.keep.uiv.p1.rcr-stratified@0.1.keepN\endcsname{0}
\expandafter\def\csname pct@rev.pol.keep.uiv.p1.rcr-stratified@0.1.keepN\endcsname{38.3}
\expandafter\def\csname ppt@rev.pol.keep.uiv.p1.rcr-stratified@0.1.keepN\endcsname{$+$38.3}
\expandafter\def\csname val@rev.pol.keep.uiv.p1.rcr-stratified@0.1.scoreN\endcsname{0.805}
\expandafter\def\csname cnt@rev.pol.keep.uiv.p1.rcr-stratified@0.1.scoreN\endcsname{1}
\expandafter\def\csname pct@rev.pol.keep.uiv.p1.rcr-stratified@0.1.scoreN\endcsname{80.5}
\expandafter\def\csname ppt@rev.pol.keep.uiv.p1.rcr-stratified@0.1.scoreN\endcsname{$+$80.5}
\expandafter\def\csname val@rev.pol.keep.uiv.p1.rcr-stratified@0.25.infeasible-rateN\endcsname{0.000}
\expandafter\def\csname cnt@rev.pol.keep.uiv.p1.rcr-stratified@0.25.infeasible-rateN\endcsname{0}
\expandafter\def\csname pct@rev.pol.keep.uiv.p1.rcr-stratified@0.25.infeasible-rateN\endcsname{0.0}
\expandafter\def\csname ppt@rev.pol.keep.uiv.p1.rcr-stratified@0.25.infeasible-rateN\endcsname{0.0}
\expandafter\def\csname val@rev.pol.keep.uiv.p1.rcr-stratified@0.25.keep20\endcsname{0.388}
\expandafter\def\csname cnt@rev.pol.keep.uiv.p1.rcr-stratified@0.25.keep20\endcsname{0}
\expandafter\def\csname pct@rev.pol.keep.uiv.p1.rcr-stratified@0.25.keep20\endcsname{38.8}
\expandafter\def\csname ppt@rev.pol.keep.uiv.p1.rcr-stratified@0.25.keep20\endcsname{$+$38.8}
\expandafter\def\csname val@rev.pol.keep.uiv.p1.rcr-stratified@0.25.keepN\endcsname{0.379}
\expandafter\def\csname cnt@rev.pol.keep.uiv.p1.rcr-stratified@0.25.keepN\endcsname{0}
\expandafter\def\csname pct@rev.pol.keep.uiv.p1.rcr-stratified@0.25.keepN\endcsname{37.9}
\expandafter\def\csname ppt@rev.pol.keep.uiv.p1.rcr-stratified@0.25.keepN\endcsname{$+$37.9}
\expandafter\def\csname val@rev.pol.keep.uiv.p1.rcr-stratified@0.25.scoreN\endcsname{0.805}
\expandafter\def\csname cnt@rev.pol.keep.uiv.p1.rcr-stratified@0.25.scoreN\endcsname{1}
\expandafter\def\csname pct@rev.pol.keep.uiv.p1.rcr-stratified@0.25.scoreN\endcsname{80.5}
\expandafter\def\csname ppt@rev.pol.keep.uiv.p1.rcr-stratified@0.25.scoreN\endcsname{$+$80.5}
\expandafter\def\csname val@rev.pol.keep.uiv.p1.rcr-stratified@0.5.infeasible-rateN\endcsname{0.000}
\expandafter\def\csname cnt@rev.pol.keep.uiv.p1.rcr-stratified@0.5.infeasible-rateN\endcsname{0}
\expandafter\def\csname pct@rev.pol.keep.uiv.p1.rcr-stratified@0.5.infeasible-rateN\endcsname{0.0}
\expandafter\def\csname ppt@rev.pol.keep.uiv.p1.rcr-stratified@0.5.infeasible-rateN\endcsname{0.0}
\expandafter\def\csname val@rev.pol.keep.uiv.p1.rcr-stratified@0.5.keep20\endcsname{0.463}
\expandafter\def\csname cnt@rev.pol.keep.uiv.p1.rcr-stratified@0.5.keep20\endcsname{0}
\expandafter\def\csname pct@rev.pol.keep.uiv.p1.rcr-stratified@0.5.keep20\endcsname{46.2}
\expandafter\def\csname ppt@rev.pol.keep.uiv.p1.rcr-stratified@0.5.keep20\endcsname{$+$46.2}
\expandafter\def\csname val@rev.pol.keep.uiv.p1.rcr-stratified@0.5.keepN\endcsname{0.413}
\expandafter\def\csname cnt@rev.pol.keep.uiv.p1.rcr-stratified@0.5.keepN\endcsname{0}
\expandafter\def\csname pct@rev.pol.keep.uiv.p1.rcr-stratified@0.5.keepN\endcsname{41.3}
\expandafter\def\csname ppt@rev.pol.keep.uiv.p1.rcr-stratified@0.5.keepN\endcsname{$+$41.3}
\expandafter\def\csname val@rev.pol.keep.uiv.p1.rcr-stratified@0.5.scoreN\endcsname{0.804}
\expandafter\def\csname cnt@rev.pol.keep.uiv.p1.rcr-stratified@0.5.scoreN\endcsname{1}
\expandafter\def\csname pct@rev.pol.keep.uiv.p1.rcr-stratified@0.5.scoreN\endcsname{80.4}
\expandafter\def\csname ppt@rev.pol.keep.uiv.p1.rcr-stratified@0.5.scoreN\endcsname{$+$80.4}
\expandafter\def\csname val@rev.pol.keep.uiv.p1.rcr-stratified@1.0.infeasible-rateN\endcsname{0.000}
\expandafter\def\csname cnt@rev.pol.keep.uiv.p1.rcr-stratified@1.0.infeasible-rateN\endcsname{0}
\expandafter\def\csname pct@rev.pol.keep.uiv.p1.rcr-stratified@1.0.infeasible-rateN\endcsname{0.0}
\expandafter\def\csname ppt@rev.pol.keep.uiv.p1.rcr-stratified@1.0.infeasible-rateN\endcsname{0.0}
\expandafter\def\csname val@rev.pol.keep.uiv.p1.rcr-stratified@1.0.keep20\endcsname{0.500}
\expandafter\def\csname cnt@rev.pol.keep.uiv.p1.rcr-stratified@1.0.keep20\endcsname{0}
\expandafter\def\csname pct@rev.pol.keep.uiv.p1.rcr-stratified@1.0.keep20\endcsname{50.0}
\expandafter\def\csname ppt@rev.pol.keep.uiv.p1.rcr-stratified@1.0.keep20\endcsname{$+$50.0}
\expandafter\def\csname val@rev.pol.keep.uiv.p1.rcr-stratified@1.0.keepN\endcsname{0.500}
\expandafter\def\csname cnt@rev.pol.keep.uiv.p1.rcr-stratified@1.0.keepN\endcsname{0}
\expandafter\def\csname pct@rev.pol.keep.uiv.p1.rcr-stratified@1.0.keepN\endcsname{50.0}
\expandafter\def\csname ppt@rev.pol.keep.uiv.p1.rcr-stratified@1.0.keepN\endcsname{$+$50.0}
\expandafter\def\csname val@rev.pol.keep.uiv.p1.rcr-stratified@1.0.scoreN\endcsname{0.793}
\expandafter\def\csname cnt@rev.pol.keep.uiv.p1.rcr-stratified@1.0.scoreN\endcsname{1}
\expandafter\def\csname pct@rev.pol.keep.uiv.p1.rcr-stratified@1.0.scoreN\endcsname{79.3}
\expandafter\def\csname ppt@rev.pol.keep.uiv.p1.rcr-stratified@1.0.scoreN\endcsname{$+$79.3}
\expandafter\def\csname val@rev.pol.keep.uiv.p1.replay-only.keep20\endcsname{0.000}
\expandafter\def\csname cnt@rev.pol.keep.uiv.p1.replay-only.keep20\endcsname{0}
\expandafter\def\csname pct@rev.pol.keep.uiv.p1.replay-only.keep20\endcsname{0.0}
\expandafter\def\csname ppt@rev.pol.keep.uiv.p1.replay-only.keep20\endcsname{0.0}
\expandafter\def\csname val@rev.pol.keep.uiv.p1.replay-only.keepN\endcsname{0.000}
\expandafter\def\csname cnt@rev.pol.keep.uiv.p1.replay-only.keepN\endcsname{0}
\expandafter\def\csname pct@rev.pol.keep.uiv.p1.replay-only.keepN\endcsname{0.0}
\expandafter\def\csname ppt@rev.pol.keep.uiv.p1.replay-only.keepN\endcsname{0.0}
\expandafter\def\csname val@rev.pol.keep.uiv.p1.replay-only.scoreN\endcsname{0.833}
\expandafter\def\csname cnt@rev.pol.keep.uiv.p1.replay-only.scoreN\endcsname{1}
\expandafter\def\csname pct@rev.pol.keep.uiv.p1.replay-only.scoreN\endcsname{83.3}
\expandafter\def\csname ppt@rev.pol.keep.uiv.p1.replay-only.scoreN\endcsname{$+$83.3}
\expandafter\def\csname val@rev.pol.keep.uiv.p1.replay-ppi-uniform@0.1.infeasible-rateN\endcsname{0.000}
\expandafter\def\csname cnt@rev.pol.keep.uiv.p1.replay-ppi-uniform@0.1.infeasible-rateN\endcsname{0}
\expandafter\def\csname pct@rev.pol.keep.uiv.p1.replay-ppi-uniform@0.1.infeasible-rateN\endcsname{0.0}
\expandafter\def\csname ppt@rev.pol.keep.uiv.p1.replay-ppi-uniform@0.1.infeasible-rateN\endcsname{0.0}
\expandafter\def\csname val@rev.pol.keep.uiv.p1.replay-ppi-uniform@0.1.keep20\endcsname{0.400}
\expandafter\def\csname cnt@rev.pol.keep.uiv.p1.replay-ppi-uniform@0.1.keep20\endcsname{0}
\expandafter\def\csname pct@rev.pol.keep.uiv.p1.replay-ppi-uniform@0.1.keep20\endcsname{40.0}
\expandafter\def\csname ppt@rev.pol.keep.uiv.p1.replay-ppi-uniform@0.1.keep20\endcsname{$+$40.0}
\expandafter\def\csname val@rev.pol.keep.uiv.p1.replay-ppi-uniform@0.1.keepN\endcsname{0.430}
\expandafter\def\csname cnt@rev.pol.keep.uiv.p1.replay-ppi-uniform@0.1.keepN\endcsname{0}
\expandafter\def\csname pct@rev.pol.keep.uiv.p1.replay-ppi-uniform@0.1.keepN\endcsname{43.0}
\expandafter\def\csname ppt@rev.pol.keep.uiv.p1.replay-ppi-uniform@0.1.keepN\endcsname{$+$43.0}
\expandafter\def\csname val@rev.pol.keep.uiv.p1.replay-ppi-uniform@0.1.scoreN\endcsname{0.801}
\expandafter\def\csname cnt@rev.pol.keep.uiv.p1.replay-ppi-uniform@0.1.scoreN\endcsname{1}
\expandafter\def\csname pct@rev.pol.keep.uiv.p1.replay-ppi-uniform@0.1.scoreN\endcsname{80.1}
\expandafter\def\csname ppt@rev.pol.keep.uiv.p1.replay-ppi-uniform@0.1.scoreN\endcsname{$+$80.1}
\expandafter\def\csname val@rev.pol.keep.uiv.p1.replay-ppi-uniform@0.25.infeasible-rateN\endcsname{0.000}
\expandafter\def\csname cnt@rev.pol.keep.uiv.p1.replay-ppi-uniform@0.25.infeasible-rateN\endcsname{0}
\expandafter\def\csname pct@rev.pol.keep.uiv.p1.replay-ppi-uniform@0.25.infeasible-rateN\endcsname{0.0}
\expandafter\def\csname ppt@rev.pol.keep.uiv.p1.replay-ppi-uniform@0.25.infeasible-rateN\endcsname{0.0}
\expandafter\def\csname val@rev.pol.keep.uiv.p1.replay-ppi-uniform@0.25.keep20\endcsname{0.412}
\expandafter\def\csname cnt@rev.pol.keep.uiv.p1.replay-ppi-uniform@0.25.keep20\endcsname{0}
\expandafter\def\csname pct@rev.pol.keep.uiv.p1.replay-ppi-uniform@0.25.keep20\endcsname{41.2}
\expandafter\def\csname ppt@rev.pol.keep.uiv.p1.replay-ppi-uniform@0.25.keep20\endcsname{$+$41.2}
\expandafter\def\csname val@rev.pol.keep.uiv.p1.replay-ppi-uniform@0.25.keepN\endcsname{0.435}
\expandafter\def\csname cnt@rev.pol.keep.uiv.p1.replay-ppi-uniform@0.25.keepN\endcsname{0}
\expandafter\def\csname pct@rev.pol.keep.uiv.p1.replay-ppi-uniform@0.25.keepN\endcsname{43.5}
\expandafter\def\csname ppt@rev.pol.keep.uiv.p1.replay-ppi-uniform@0.25.keepN\endcsname{$+$43.5}
\expandafter\def\csname val@rev.pol.keep.uiv.p1.replay-ppi-uniform@0.25.scoreN\endcsname{0.800}
\expandafter\def\csname cnt@rev.pol.keep.uiv.p1.replay-ppi-uniform@0.25.scoreN\endcsname{1}
\expandafter\def\csname pct@rev.pol.keep.uiv.p1.replay-ppi-uniform@0.25.scoreN\endcsname{80.0}
\expandafter\def\csname ppt@rev.pol.keep.uiv.p1.replay-ppi-uniform@0.25.scoreN\endcsname{$+$80.0}
\expandafter\def\csname val@rev.pol.keep.uiv.p1.replay-ppi-uniform@0.5.infeasible-rateN\endcsname{0.000}
\expandafter\def\csname cnt@rev.pol.keep.uiv.p1.replay-ppi-uniform@0.5.infeasible-rateN\endcsname{0}
\expandafter\def\csname pct@rev.pol.keep.uiv.p1.replay-ppi-uniform@0.5.infeasible-rateN\endcsname{0.0}
\expandafter\def\csname ppt@rev.pol.keep.uiv.p1.replay-ppi-uniform@0.5.infeasible-rateN\endcsname{0.0}
\expandafter\def\csname val@rev.pol.keep.uiv.p1.replay-ppi-uniform@0.5.keep20\endcsname{0.325}
\expandafter\def\csname cnt@rev.pol.keep.uiv.p1.replay-ppi-uniform@0.5.keep20\endcsname{0}
\expandafter\def\csname pct@rev.pol.keep.uiv.p1.replay-ppi-uniform@0.5.keep20\endcsname{32.5}
\expandafter\def\csname ppt@rev.pol.keep.uiv.p1.replay-ppi-uniform@0.5.keep20\endcsname{$+$32.5}
\expandafter\def\csname val@rev.pol.keep.uiv.p1.replay-ppi-uniform@0.5.keepN\endcsname{0.414}
\expandafter\def\csname cnt@rev.pol.keep.uiv.p1.replay-ppi-uniform@0.5.keepN\endcsname{0}
\expandafter\def\csname pct@rev.pol.keep.uiv.p1.replay-ppi-uniform@0.5.keepN\endcsname{41.4}
\expandafter\def\csname ppt@rev.pol.keep.uiv.p1.replay-ppi-uniform@0.5.keepN\endcsname{$+$41.4}
\expandafter\def\csname val@rev.pol.keep.uiv.p1.replay-ppi-uniform@0.5.scoreN\endcsname{0.803}
\expandafter\def\csname cnt@rev.pol.keep.uiv.p1.replay-ppi-uniform@0.5.scoreN\endcsname{1}
\expandafter\def\csname pct@rev.pol.keep.uiv.p1.replay-ppi-uniform@0.5.scoreN\endcsname{80.3}
\expandafter\def\csname ppt@rev.pol.keep.uiv.p1.replay-ppi-uniform@0.5.scoreN\endcsname{$+$80.3}
\expandafter\def\csname val@rev.pol.keep.uiv.p1.replay-ppi-uniform@1.0.infeasible-rateN\endcsname{0.000}
\expandafter\def\csname cnt@rev.pol.keep.uiv.p1.replay-ppi-uniform@1.0.infeasible-rateN\endcsname{0}
\expandafter\def\csname pct@rev.pol.keep.uiv.p1.replay-ppi-uniform@1.0.infeasible-rateN\endcsname{0.0}
\expandafter\def\csname ppt@rev.pol.keep.uiv.p1.replay-ppi-uniform@1.0.infeasible-rateN\endcsname{0.0}
\expandafter\def\csname val@rev.pol.keep.uiv.p1.replay-ppi-uniform@1.0.keep20\endcsname{0.500}
\expandafter\def\csname cnt@rev.pol.keep.uiv.p1.replay-ppi-uniform@1.0.keep20\endcsname{0}
\expandafter\def\csname pct@rev.pol.keep.uiv.p1.replay-ppi-uniform@1.0.keep20\endcsname{50.0}
\expandafter\def\csname ppt@rev.pol.keep.uiv.p1.replay-ppi-uniform@1.0.keep20\endcsname{$+$50.0}
\expandafter\def\csname val@rev.pol.keep.uiv.p1.replay-ppi-uniform@1.0.keepN\endcsname{0.500}
\expandafter\def\csname cnt@rev.pol.keep.uiv.p1.replay-ppi-uniform@1.0.keepN\endcsname{0}
\expandafter\def\csname pct@rev.pol.keep.uiv.p1.replay-ppi-uniform@1.0.keepN\endcsname{50.0}
\expandafter\def\csname ppt@rev.pol.keep.uiv.p1.replay-ppi-uniform@1.0.keepN\endcsname{$+$50.0}
\expandafter\def\csname val@rev.pol.keep.uiv.p1.replay-ppi-uniform@1.0.scoreN\endcsname{0.793}
\expandafter\def\csname cnt@rev.pol.keep.uiv.p1.replay-ppi-uniform@1.0.scoreN\endcsname{1}
\expandafter\def\csname pct@rev.pol.keep.uiv.p1.replay-ppi-uniform@1.0.scoreN\endcsname{79.3}
\expandafter\def\csname ppt@rev.pol.keep.uiv.p1.replay-ppi-uniform@1.0.scoreN\endcsname{$+$79.3}
\expandafter\def\csname val@rev.pol.keep.uiv.p1.suspend-all.keep20\endcsname{0.000}
\expandafter\def\csname cnt@rev.pol.keep.uiv.p1.suspend-all.keep20\endcsname{0}
\expandafter\def\csname pct@rev.pol.keep.uiv.p1.suspend-all.keep20\endcsname{0.0}
\expandafter\def\csname ppt@rev.pol.keep.uiv.p1.suspend-all.keep20\endcsname{0.0}
\expandafter\def\csname val@rev.pol.keep.uiv.p1.suspend-all.keepN\endcsname{0.000}
\expandafter\def\csname cnt@rev.pol.keep.uiv.p1.suspend-all.keepN\endcsname{0}
\expandafter\def\csname pct@rev.pol.keep.uiv.p1.suspend-all.keepN\endcsname{0.0}
\expandafter\def\csname ppt@rev.pol.keep.uiv.p1.suspend-all.keepN\endcsname{0.0}
\expandafter\def\csname val@rev.pol.keep.uiv.p1.suspend-all.scoreN\endcsname{0.833}
\expandafter\def\csname cnt@rev.pol.keep.uiv.p1.suspend-all.scoreN\endcsname{1}
\expandafter\def\csname pct@rev.pol.keep.uiv.p1.suspend-all.scoreN\endcsname{83.3}
\expandafter\def\csname ppt@rev.pol.keep.uiv.p1.suspend-all.scoreN\endcsname{$+$83.3}
\expandafter\def\csname val@rev.pol.keep.uiv.p1.uniform-fresh@0.1.infeasible-rateN\endcsname{0.000}
\expandafter\def\csname cnt@rev.pol.keep.uiv.p1.uniform-fresh@0.1.infeasible-rateN\endcsname{0}
\expandafter\def\csname pct@rev.pol.keep.uiv.p1.uniform-fresh@0.1.infeasible-rateN\endcsname{0.0}
\expandafter\def\csname ppt@rev.pol.keep.uiv.p1.uniform-fresh@0.1.infeasible-rateN\endcsname{0.0}
\expandafter\def\csname val@rev.pol.keep.uiv.p1.uniform-fresh@0.1.keep20\endcsname{0.588}
\expandafter\def\csname cnt@rev.pol.keep.uiv.p1.uniform-fresh@0.1.keep20\endcsname{1}
\expandafter\def\csname pct@rev.pol.keep.uiv.p1.uniform-fresh@0.1.keep20\endcsname{58.8}
\expandafter\def\csname ppt@rev.pol.keep.uiv.p1.uniform-fresh@0.1.keep20\endcsname{$+$58.8}
\expandafter\def\csname val@rev.pol.keep.uiv.p1.uniform-fresh@0.1.keepN\endcsname{0.668}
\expandafter\def\csname cnt@rev.pol.keep.uiv.p1.uniform-fresh@0.1.keepN\endcsname{1}
\expandafter\def\csname pct@rev.pol.keep.uiv.p1.uniform-fresh@0.1.keepN\endcsname{66.8}
\expandafter\def\csname ppt@rev.pol.keep.uiv.p1.uniform-fresh@0.1.keepN\endcsname{$+$66.8}
\expandafter\def\csname val@rev.pol.keep.uiv.p1.uniform-fresh@0.1.scoreN\endcsname{0.785}
\expandafter\def\csname cnt@rev.pol.keep.uiv.p1.uniform-fresh@0.1.scoreN\endcsname{1}
\expandafter\def\csname pct@rev.pol.keep.uiv.p1.uniform-fresh@0.1.scoreN\endcsname{78.5}
\expandafter\def\csname ppt@rev.pol.keep.uiv.p1.uniform-fresh@0.1.scoreN\endcsname{$+$78.5}
\expandafter\def\csname val@rev.pol.keep.uiv.p1.uniform-fresh@0.25.infeasible-rateN\endcsname{0.000}
\expandafter\def\csname cnt@rev.pol.keep.uiv.p1.uniform-fresh@0.25.infeasible-rateN\endcsname{0}
\expandafter\def\csname pct@rev.pol.keep.uiv.p1.uniform-fresh@0.25.infeasible-rateN\endcsname{0.0}
\expandafter\def\csname ppt@rev.pol.keep.uiv.p1.uniform-fresh@0.25.infeasible-rateN\endcsname{0.0}
\expandafter\def\csname val@rev.pol.keep.uiv.p1.uniform-fresh@0.25.keep20\endcsname{0.500}
\expandafter\def\csname cnt@rev.pol.keep.uiv.p1.uniform-fresh@0.25.keep20\endcsname{0}
\expandafter\def\csname pct@rev.pol.keep.uiv.p1.uniform-fresh@0.25.keep20\endcsname{50.0}
\expandafter\def\csname ppt@rev.pol.keep.uiv.p1.uniform-fresh@0.25.keep20\endcsname{$+$50.0}
\expandafter\def\csname val@rev.pol.keep.uiv.p1.uniform-fresh@0.25.keepN\endcsname{0.570}
\expandafter\def\csname cnt@rev.pol.keep.uiv.p1.uniform-fresh@0.25.keepN\endcsname{1}
\expandafter\def\csname pct@rev.pol.keep.uiv.p1.uniform-fresh@0.25.keepN\endcsname{57.0}
\expandafter\def\csname ppt@rev.pol.keep.uiv.p1.uniform-fresh@0.25.keepN\endcsname{$+$57.0}
\expandafter\def\csname val@rev.pol.keep.uiv.p1.uniform-fresh@0.25.scoreN\endcsname{0.792}
\expandafter\def\csname cnt@rev.pol.keep.uiv.p1.uniform-fresh@0.25.scoreN\endcsname{1}
\expandafter\def\csname pct@rev.pol.keep.uiv.p1.uniform-fresh@0.25.scoreN\endcsname{79.2}
\expandafter\def\csname ppt@rev.pol.keep.uiv.p1.uniform-fresh@0.25.scoreN\endcsname{$+$79.2}
\expandafter\def\csname val@rev.pol.keep.uiv.p1.uniform-fresh@0.5.infeasible-rateN\endcsname{0.000}
\expandafter\def\csname cnt@rev.pol.keep.uiv.p1.uniform-fresh@0.5.infeasible-rateN\endcsname{0}
\expandafter\def\csname pct@rev.pol.keep.uiv.p1.uniform-fresh@0.5.infeasible-rateN\endcsname{0.0}
\expandafter\def\csname ppt@rev.pol.keep.uiv.p1.uniform-fresh@0.5.infeasible-rateN\endcsname{0.0}
\expandafter\def\csname val@rev.pol.keep.uiv.p1.uniform-fresh@0.5.keep20\endcsname{0.388}
\expandafter\def\csname cnt@rev.pol.keep.uiv.p1.uniform-fresh@0.5.keep20\endcsname{0}
\expandafter\def\csname pct@rev.pol.keep.uiv.p1.uniform-fresh@0.5.keep20\endcsname{38.8}
\expandafter\def\csname ppt@rev.pol.keep.uiv.p1.uniform-fresh@0.5.keep20\endcsname{$+$38.8}
\expandafter\def\csname val@rev.pol.keep.uiv.p1.uniform-fresh@0.5.keepN\endcsname{0.504}
\expandafter\def\csname cnt@rev.pol.keep.uiv.p1.uniform-fresh@0.5.keepN\endcsname{1}
\expandafter\def\csname pct@rev.pol.keep.uiv.p1.uniform-fresh@0.5.keepN\endcsname{50.4}
\expandafter\def\csname ppt@rev.pol.keep.uiv.p1.uniform-fresh@0.5.keepN\endcsname{$+$50.4}
\expandafter\def\csname val@rev.pol.keep.uiv.p1.uniform-fresh@0.5.scoreN\endcsname{0.796}
\expandafter\def\csname cnt@rev.pol.keep.uiv.p1.uniform-fresh@0.5.scoreN\endcsname{1}
\expandafter\def\csname pct@rev.pol.keep.uiv.p1.uniform-fresh@0.5.scoreN\endcsname{79.6}
\expandafter\def\csname ppt@rev.pol.keep.uiv.p1.uniform-fresh@0.5.scoreN\endcsname{$+$79.6}
\expandafter\def\csname val@rev.pol.keep.uiv.p1.uniform-fresh@1.0.infeasible-rateN\endcsname{0.000}
\expandafter\def\csname cnt@rev.pol.keep.uiv.p1.uniform-fresh@1.0.infeasible-rateN\endcsname{0}
\expandafter\def\csname pct@rev.pol.keep.uiv.p1.uniform-fresh@1.0.infeasible-rateN\endcsname{0.0}
\expandafter\def\csname ppt@rev.pol.keep.uiv.p1.uniform-fresh@1.0.infeasible-rateN\endcsname{0.0}
\expandafter\def\csname val@rev.pol.keep.uiv.p1.uniform-fresh@1.0.keep20\endcsname{0.500}
\expandafter\def\csname cnt@rev.pol.keep.uiv.p1.uniform-fresh@1.0.keep20\endcsname{0}
\expandafter\def\csname pct@rev.pol.keep.uiv.p1.uniform-fresh@1.0.keep20\endcsname{50.0}
\expandafter\def\csname ppt@rev.pol.keep.uiv.p1.uniform-fresh@1.0.keep20\endcsname{$+$50.0}
\expandafter\def\csname val@rev.pol.keep.uiv.p1.uniform-fresh@1.0.keepN\endcsname{0.500}
\expandafter\def\csname cnt@rev.pol.keep.uiv.p1.uniform-fresh@1.0.keepN\endcsname{0}
\expandafter\def\csname pct@rev.pol.keep.uiv.p1.uniform-fresh@1.0.keepN\endcsname{50.0}
\expandafter\def\csname ppt@rev.pol.keep.uiv.p1.uniform-fresh@1.0.keepN\endcsname{$+$50.0}
\expandafter\def\csname val@rev.pol.keep.uiv.p1.uniform-fresh@1.0.scoreN\endcsname{0.793}
\expandafter\def\csname cnt@rev.pol.keep.uiv.p1.uniform-fresh@1.0.scoreN\endcsname{1}
\expandafter\def\csname pct@rev.pol.keep.uiv.p1.uniform-fresh@1.0.scoreN\endcsname{79.3}
\expandafter\def\csname ppt@rev.pol.keep.uiv.p1.uniform-fresh@1.0.scoreN\endcsname{$+$79.3}
\expandafter\def\csname val@rev.pol.keep.uiv.p2.best-single@0.1.share-direct-N\endcsname{0.490}
\expandafter\def\csname cnt@rev.pol.keep.uiv.p2.best-single@0.1.share-direct-N\endcsname{0}
\expandafter\def\csname pct@rev.pol.keep.uiv.p2.best-single@0.1.share-direct-N\endcsname{49.0}
\expandafter\def\csname ppt@rev.pol.keep.uiv.p2.best-single@0.1.share-direct-N\endcsname{$+$49.0}
\expandafter\def\csname val@rev.pol.keep.uiv.p2.best-single@0.25.share-direct-N\endcsname{0.373}
\expandafter\def\csname cnt@rev.pol.keep.uiv.p2.best-single@0.25.share-direct-N\endcsname{0}
\expandafter\def\csname pct@rev.pol.keep.uiv.p2.best-single@0.25.share-direct-N\endcsname{37.3}
\expandafter\def\csname ppt@rev.pol.keep.uiv.p2.best-single@0.25.share-direct-N\endcsname{$+$37.3}
\expandafter\def\csname val@rev.pol.keep.uiv.p2.best-single@0.5.share-direct-N\endcsname{0.308}
\expandafter\def\csname cnt@rev.pol.keep.uiv.p2.best-single@0.5.share-direct-N\endcsname{0}
\expandafter\def\csname pct@rev.pol.keep.uiv.p2.best-single@0.5.share-direct-N\endcsname{30.9}
\expandafter\def\csname ppt@rev.pol.keep.uiv.p2.best-single@0.5.share-direct-N\endcsname{$+$30.9}
\expandafter\def\csname val@rev.pol.keep.uiv.p2.best-single@1.0.share-direct-N\endcsname{0.000}
\expandafter\def\csname cnt@rev.pol.keep.uiv.p2.best-single@1.0.share-direct-N\endcsname{0}
\expandafter\def\csname pct@rev.pol.keep.uiv.p2.best-single@1.0.share-direct-N\endcsname{0.0}
\expandafter\def\csname ppt@rev.pol.keep.uiv.p2.best-single@1.0.share-direct-N\endcsname{0.0}
\expandafter\def\csname val@rev.pol.keep.uiv.p2.corr.budgeted-points\endcsname{16}
\expandafter\def\csname cnt@rev.pol.keep.uiv.p2.corr.budgeted-points\endcsname{16}
\expandafter\def\csname pct@rev.pol.keep.uiv.p2.corr.budgeted-points\endcsname{1600.0}
\expandafter\def\csname ppt@rev.pol.keep.uiv.p2.corr.budgeted-points\endcsname{$+$1600.0}
\expandafter\def\csname val@rev.pol.keep.uiv.p2.corr.pearson\endcsname{-0.077}
\expandafter\def\csname cnt@rev.pol.keep.uiv.p2.corr.pearson\endcsname{0}
\expandafter\def\csname pct@rev.pol.keep.uiv.p2.corr.pearson\endcsname{$-$7.7}
\expandafter\def\csname ppt@rev.pol.keep.uiv.p2.corr.pearson\endcsname{$-$7.7}
\expandafter\def\csname val@rev.pol.keep.uiv.p2.corr.pearson-20\endcsname{-0.222}
\expandafter\def\csname cnt@rev.pol.keep.uiv.p2.corr.pearson-20\endcsname{0}
\expandafter\def\csname pct@rev.pol.keep.uiv.p2.corr.pearson-20\endcsname{$-$22.2}
\expandafter\def\csname ppt@rev.pol.keep.uiv.p2.corr.pearson-20\endcsname{$-$22.2}
\expandafter\def\csname val@rev.pol.keep.uiv.p2.corr.pearson-all\endcsname{-0.560}
\expandafter\def\csname cnt@rev.pol.keep.uiv.p2.corr.pearson-all\endcsname{-1}
\expandafter\def\csname pct@rev.pol.keep.uiv.p2.corr.pearson-all\endcsname{$-$56.0}
\expandafter\def\csname ppt@rev.pol.keep.uiv.p2.corr.pearson-all\endcsname{$-$56.0}
\expandafter\def\csname val@rev.pol.keep.uiv.p2.corr.slope-pt-per-keep\endcsname{-0.002}
\expandafter\def\csname cnt@rev.pol.keep.uiv.p2.corr.slope-pt-per-keep\endcsname{0}
\expandafter\def\csname pct@rev.pol.keep.uiv.p2.corr.slope-pt-per-keep\endcsname{$-$0.2}
\expandafter\def\csname ppt@rev.pol.keep.uiv.p2.corr.slope-pt-per-keep\endcsname{$-$0.2}
\expandafter\def\csname val@rev.pol.keep.uiv.p2.corr.spearman\endcsname{0.110}
\expandafter\def\csname cnt@rev.pol.keep.uiv.p2.corr.spearman\endcsname{0}
\expandafter\def\csname pct@rev.pol.keep.uiv.p2.corr.spearman\endcsname{11.0}
\expandafter\def\csname ppt@rev.pol.keep.uiv.p2.corr.spearman\endcsname{$+$11.0}
\expandafter\def\csname val@rev.pol.keep.uiv.p2.corr.spearman-20\endcsname{-0.039}
\expandafter\def\csname cnt@rev.pol.keep.uiv.p2.corr.spearman-20\endcsname{0}
\expandafter\def\csname pct@rev.pol.keep.uiv.p2.corr.spearman-20\endcsname{$-$3.9}
\expandafter\def\csname ppt@rev.pol.keep.uiv.p2.corr.spearman-20\endcsname{$-$3.9}
\expandafter\def\csname val@rev.pol.keep.uiv.p2.corr.spearman-all\endcsname{-0.161}
\expandafter\def\csname cnt@rev.pol.keep.uiv.p2.corr.spearman-all\endcsname{0}
\expandafter\def\csname pct@rev.pol.keep.uiv.p2.corr.spearman-all\endcsname{$-$16.1}
\expandafter\def\csname ppt@rev.pol.keep.uiv.p2.corr.spearman-all\endcsname{$-$16.1}
\expandafter\def\csname val@rev.pol.keep.uiv.p2.corr.spearman-primary\endcsname{-0.200}
\expandafter\def\csname cnt@rev.pol.keep.uiv.p2.corr.spearman-primary\endcsname{0}
\expandafter\def\csname pct@rev.pol.keep.uiv.p2.corr.spearman-primary\endcsname{$-$20.0}
\expandafter\def\csname ppt@rev.pol.keep.uiv.p2.corr.spearman-primary\endcsname{$-$20.0}
\expandafter\def\csname val@rev.pol.keep.uiv.p2.corr.with-nobudget-points\endcsname{21}
\expandafter\def\csname cnt@rev.pol.keep.uiv.p2.corr.with-nobudget-points\endcsname{21}
\expandafter\def\csname pct@rev.pol.keep.uiv.p2.corr.with-nobudget-points\endcsname{2100.0}
\expandafter\def\csname ppt@rev.pol.keep.uiv.p2.corr.with-nobudget-points\endcsname{$+$2100.0}
\expandafter\def\csname val@rev.pol.keep.uiv.p2.full-recert.keep20\endcsname{0.750}
\expandafter\def\csname cnt@rev.pol.keep.uiv.p2.full-recert.keep20\endcsname{1}
\expandafter\def\csname pct@rev.pol.keep.uiv.p2.full-recert.keep20\endcsname{75.0}
\expandafter\def\csname ppt@rev.pol.keep.uiv.p2.full-recert.keep20\endcsname{$+$75.0}
\expandafter\def\csname val@rev.pol.keep.uiv.p2.full-recert.keepN\endcsname{0.750}
\expandafter\def\csname cnt@rev.pol.keep.uiv.p2.full-recert.keepN\endcsname{1}
\expandafter\def\csname pct@rev.pol.keep.uiv.p2.full-recert.keepN\endcsname{75.0}
\expandafter\def\csname ppt@rev.pol.keep.uiv.p2.full-recert.keepN\endcsname{$+$75.0}
\expandafter\def\csname val@rev.pol.keep.uiv.p2.full-recert.scoreN\endcsname{0.832}
\expandafter\def\csname cnt@rev.pol.keep.uiv.p2.full-recert.scoreN\endcsname{1}
\expandafter\def\csname pct@rev.pol.keep.uiv.p2.full-recert.scoreN\endcsname{83.2}
\expandafter\def\csname ppt@rev.pol.keep.uiv.p2.full-recert.scoreN\endcsname{$+$83.2}
\expandafter\def\csname val@rev.pol.keep.uiv.p2.generic-ppi-source@0.1.infeasible-rateN\endcsname{0.000}
\expandafter\def\csname cnt@rev.pol.keep.uiv.p2.generic-ppi-source@0.1.infeasible-rateN\endcsname{0}
\expandafter\def\csname pct@rev.pol.keep.uiv.p2.generic-ppi-source@0.1.infeasible-rateN\endcsname{0.0}
\expandafter\def\csname ppt@rev.pol.keep.uiv.p2.generic-ppi-source@0.1.infeasible-rateN\endcsname{0.0}
\expandafter\def\csname val@rev.pol.keep.uiv.p2.generic-ppi-source@0.1.keep20\endcsname{0.525}
\expandafter\def\csname cnt@rev.pol.keep.uiv.p2.generic-ppi-source@0.1.keep20\endcsname{1}
\expandafter\def\csname pct@rev.pol.keep.uiv.p2.generic-ppi-source@0.1.keep20\endcsname{52.5}
\expandafter\def\csname ppt@rev.pol.keep.uiv.p2.generic-ppi-source@0.1.keep20\endcsname{$+$52.5}
\expandafter\def\csname val@rev.pol.keep.uiv.p2.generic-ppi-source@0.1.keepN\endcsname{0.461}
\expandafter\def\csname cnt@rev.pol.keep.uiv.p2.generic-ppi-source@0.1.keepN\endcsname{0}
\expandafter\def\csname pct@rev.pol.keep.uiv.p2.generic-ppi-source@0.1.keepN\endcsname{46.1}
\expandafter\def\csname ppt@rev.pol.keep.uiv.p2.generic-ppi-source@0.1.keepN\endcsname{$+$46.1}
\expandafter\def\csname val@rev.pol.keep.uiv.p2.generic-ppi-source@0.1.scoreN\endcsname{0.831}
\expandafter\def\csname cnt@rev.pol.keep.uiv.p2.generic-ppi-source@0.1.scoreN\endcsname{1}
\expandafter\def\csname pct@rev.pol.keep.uiv.p2.generic-ppi-source@0.1.scoreN\endcsname{83.1}
\expandafter\def\csname ppt@rev.pol.keep.uiv.p2.generic-ppi-source@0.1.scoreN\endcsname{$+$83.1}
\expandafter\def\csname val@rev.pol.keep.uiv.p2.generic-ppi-source@0.25.infeasible-rateN\endcsname{0.000}
\expandafter\def\csname cnt@rev.pol.keep.uiv.p2.generic-ppi-source@0.25.infeasible-rateN\endcsname{0}
\expandafter\def\csname pct@rev.pol.keep.uiv.p2.generic-ppi-source@0.25.infeasible-rateN\endcsname{0.0}
\expandafter\def\csname ppt@rev.pol.keep.uiv.p2.generic-ppi-source@0.25.infeasible-rateN\endcsname{0.0}
\expandafter\def\csname val@rev.pol.keep.uiv.p2.generic-ppi-source@0.25.keep20\endcsname{0.550}
\expandafter\def\csname cnt@rev.pol.keep.uiv.p2.generic-ppi-source@0.25.keep20\endcsname{1}
\expandafter\def\csname pct@rev.pol.keep.uiv.p2.generic-ppi-source@0.25.keep20\endcsname{55.0}
\expandafter\def\csname ppt@rev.pol.keep.uiv.p2.generic-ppi-source@0.25.keep20\endcsname{$+$55.0}
\expandafter\def\csname val@rev.pol.keep.uiv.p2.generic-ppi-source@0.25.keepN\endcsname{0.544}
\expandafter\def\csname cnt@rev.pol.keep.uiv.p2.generic-ppi-source@0.25.keepN\endcsname{1}
\expandafter\def\csname pct@rev.pol.keep.uiv.p2.generic-ppi-source@0.25.keepN\endcsname{54.4}
\expandafter\def\csname ppt@rev.pol.keep.uiv.p2.generic-ppi-source@0.25.keepN\endcsname{$+$54.4}
\expandafter\def\csname val@rev.pol.keep.uiv.p2.generic-ppi-source@0.25.scoreN\endcsname{0.827}
\expandafter\def\csname cnt@rev.pol.keep.uiv.p2.generic-ppi-source@0.25.scoreN\endcsname{1}
\expandafter\def\csname pct@rev.pol.keep.uiv.p2.generic-ppi-source@0.25.scoreN\endcsname{82.7}
\expandafter\def\csname ppt@rev.pol.keep.uiv.p2.generic-ppi-source@0.25.scoreN\endcsname{$+$82.7}
\expandafter\def\csname val@rev.pol.keep.uiv.p2.generic-ppi-source@0.5.infeasible-rateN\endcsname{0.000}
\expandafter\def\csname cnt@rev.pol.keep.uiv.p2.generic-ppi-source@0.5.infeasible-rateN\endcsname{0}
\expandafter\def\csname pct@rev.pol.keep.uiv.p2.generic-ppi-source@0.5.infeasible-rateN\endcsname{0.0}
\expandafter\def\csname ppt@rev.pol.keep.uiv.p2.generic-ppi-source@0.5.infeasible-rateN\endcsname{0.0}
\expandafter\def\csname val@rev.pol.keep.uiv.p2.generic-ppi-source@0.5.keep20\endcsname{0.662}
\expandafter\def\csname cnt@rev.pol.keep.uiv.p2.generic-ppi-source@0.5.keep20\endcsname{1}
\expandafter\def\csname pct@rev.pol.keep.uiv.p2.generic-ppi-source@0.5.keep20\endcsname{66.2}
\expandafter\def\csname ppt@rev.pol.keep.uiv.p2.generic-ppi-source@0.5.keep20\endcsname{$+$66.2}
\expandafter\def\csname val@rev.pol.keep.uiv.p2.generic-ppi-source@0.5.keepN\endcsname{0.635}
\expandafter\def\csname cnt@rev.pol.keep.uiv.p2.generic-ppi-source@0.5.keepN\endcsname{1}
\expandafter\def\csname pct@rev.pol.keep.uiv.p2.generic-ppi-source@0.5.keepN\endcsname{63.5}
\expandafter\def\csname ppt@rev.pol.keep.uiv.p2.generic-ppi-source@0.5.keepN\endcsname{$+$63.5}
\expandafter\def\csname val@rev.pol.keep.uiv.p2.generic-ppi-source@0.5.scoreN\endcsname{0.827}
\expandafter\def\csname cnt@rev.pol.keep.uiv.p2.generic-ppi-source@0.5.scoreN\endcsname{1}
\expandafter\def\csname pct@rev.pol.keep.uiv.p2.generic-ppi-source@0.5.scoreN\endcsname{82.7}
\expandafter\def\csname ppt@rev.pol.keep.uiv.p2.generic-ppi-source@0.5.scoreN\endcsname{$+$82.7}
\expandafter\def\csname val@rev.pol.keep.uiv.p2.generic-ppi-source@1.0.infeasible-rateN\endcsname{0.000}
\expandafter\def\csname cnt@rev.pol.keep.uiv.p2.generic-ppi-source@1.0.infeasible-rateN\endcsname{0}
\expandafter\def\csname pct@rev.pol.keep.uiv.p2.generic-ppi-source@1.0.infeasible-rateN\endcsname{0.0}
\expandafter\def\csname ppt@rev.pol.keep.uiv.p2.generic-ppi-source@1.0.infeasible-rateN\endcsname{0.0}
\expandafter\def\csname val@rev.pol.keep.uiv.p2.generic-ppi-source@1.0.keep20\endcsname{0.750}
\expandafter\def\csname cnt@rev.pol.keep.uiv.p2.generic-ppi-source@1.0.keep20\endcsname{1}
\expandafter\def\csname pct@rev.pol.keep.uiv.p2.generic-ppi-source@1.0.keep20\endcsname{75.0}
\expandafter\def\csname ppt@rev.pol.keep.uiv.p2.generic-ppi-source@1.0.keep20\endcsname{$+$75.0}
\expandafter\def\csname val@rev.pol.keep.uiv.p2.generic-ppi-source@1.0.keepN\endcsname{0.750}
\expandafter\def\csname cnt@rev.pol.keep.uiv.p2.generic-ppi-source@1.0.keepN\endcsname{1}
\expandafter\def\csname pct@rev.pol.keep.uiv.p2.generic-ppi-source@1.0.keepN\endcsname{75.0}
\expandafter\def\csname ppt@rev.pol.keep.uiv.p2.generic-ppi-source@1.0.keepN\endcsname{$+$75.0}
\expandafter\def\csname val@rev.pol.keep.uiv.p2.generic-ppi-source@1.0.scoreN\endcsname{0.832}
\expandafter\def\csname cnt@rev.pol.keep.uiv.p2.generic-ppi-source@1.0.scoreN\endcsname{1}
\expandafter\def\csname pct@rev.pol.keep.uiv.p2.generic-ppi-source@1.0.scoreN\endcsname{83.2}
\expandafter\def\csname ppt@rev.pol.keep.uiv.p2.generic-ppi-source@1.0.scoreN\endcsname{$+$83.2}
\expandafter\def\csname val@rev.pol.keep.uiv.p2.keep-all.keep20\endcsname{1.000}
\expandafter\def\csname cnt@rev.pol.keep.uiv.p2.keep-all.keep20\endcsname{1}
\expandafter\def\csname pct@rev.pol.keep.uiv.p2.keep-all.keep20\endcsname{100.0}
\expandafter\def\csname ppt@rev.pol.keep.uiv.p2.keep-all.keep20\endcsname{$+$100.0}
\expandafter\def\csname val@rev.pol.keep.uiv.p2.keep-all.keepN\endcsname{1.000}
\expandafter\def\csname cnt@rev.pol.keep.uiv.p2.keep-all.keepN\endcsname{1}
\expandafter\def\csname pct@rev.pol.keep.uiv.p2.keep-all.keepN\endcsname{100.0}
\expandafter\def\csname ppt@rev.pol.keep.uiv.p2.keep-all.keepN\endcsname{$+$100.0}
\expandafter\def\csname val@rev.pol.keep.uiv.p2.keep-all.scoreN\endcsname{0.816}
\expandafter\def\csname cnt@rev.pol.keep.uiv.p2.keep-all.scoreN\endcsname{1}
\expandafter\def\csname pct@rev.pol.keep.uiv.p2.keep-all.scoreN\endcsname{81.6}
\expandafter\def\csname ppt@rev.pol.keep.uiv.p2.keep-all.scoreN\endcsname{$+$81.6}
\expandafter\def\csname val@rev.pol.keep.uiv.p2.old-decision.keep20\endcsname{0.500}
\expandafter\def\csname cnt@rev.pol.keep.uiv.p2.old-decision.keep20\endcsname{0}
\expandafter\def\csname pct@rev.pol.keep.uiv.p2.old-decision.keep20\endcsname{50.0}
\expandafter\def\csname ppt@rev.pol.keep.uiv.p2.old-decision.keep20\endcsname{$+$50.0}
\expandafter\def\csname val@rev.pol.keep.uiv.p2.old-decision.keepN\endcsname{0.500}
\expandafter\def\csname cnt@rev.pol.keep.uiv.p2.old-decision.keepN\endcsname{0}
\expandafter\def\csname pct@rev.pol.keep.uiv.p2.old-decision.keepN\endcsname{50.0}
\expandafter\def\csname ppt@rev.pol.keep.uiv.p2.old-decision.keepN\endcsname{$+$50.0}
\expandafter\def\csname val@rev.pol.keep.uiv.p2.old-decision.scoreN\endcsname{0.818}
\expandafter\def\csname cnt@rev.pol.keep.uiv.p2.old-decision.scoreN\endcsname{1}
\expandafter\def\csname pct@rev.pol.keep.uiv.p2.old-decision.scoreN\endcsname{81.8}
\expandafter\def\csname ppt@rev.pol.keep.uiv.p2.old-decision.scoreN\endcsname{$+$81.8}
\expandafter\def\csname val@rev.pol.keep.uiv.p2.oracle-on-T.keep20\endcsname{0.250}
\expandafter\def\csname cnt@rev.pol.keep.uiv.p2.oracle-on-T.keep20\endcsname{0}
\expandafter\def\csname pct@rev.pol.keep.uiv.p2.oracle-on-T.keep20\endcsname{25.0}
\expandafter\def\csname ppt@rev.pol.keep.uiv.p2.oracle-on-T.keep20\endcsname{$+$25.0}
\expandafter\def\csname val@rev.pol.keep.uiv.p2.oracle-on-T.keepN\endcsname{0.250}
\expandafter\def\csname cnt@rev.pol.keep.uiv.p2.oracle-on-T.keepN\endcsname{0}
\expandafter\def\csname pct@rev.pol.keep.uiv.p2.oracle-on-T.keepN\endcsname{25.0}
\expandafter\def\csname ppt@rev.pol.keep.uiv.p2.oracle-on-T.keepN\endcsname{$+$25.0}
\expandafter\def\csname val@rev.pol.keep.uiv.p2.oracle-on-T.scoreN\endcsname{0.840}
\expandafter\def\csname cnt@rev.pol.keep.uiv.p2.oracle-on-T.scoreN\endcsname{1}
\expandafter\def\csname pct@rev.pol.keep.uiv.p2.oracle-on-T.scoreN\endcsname{84.0}
\expandafter\def\csname ppt@rev.pol.keep.uiv.p2.oracle-on-T.scoreN\endcsname{$+$84.0}
\expandafter\def\csname val@rev.pol.keep.uiv.p2.rcr-stratified@0.1.infeasible-rateN\endcsname{0.000}
\expandafter\def\csname cnt@rev.pol.keep.uiv.p2.rcr-stratified@0.1.infeasible-rateN\endcsname{0}
\expandafter\def\csname pct@rev.pol.keep.uiv.p2.rcr-stratified@0.1.infeasible-rateN\endcsname{0.0}
\expandafter\def\csname ppt@rev.pol.keep.uiv.p2.rcr-stratified@0.1.infeasible-rateN\endcsname{0.0}
\expandafter\def\csname val@rev.pol.keep.uiv.p2.rcr-stratified@0.1.keep20\endcsname{0.512}
\expandafter\def\csname cnt@rev.pol.keep.uiv.p2.rcr-stratified@0.1.keep20\endcsname{1}
\expandafter\def\csname pct@rev.pol.keep.uiv.p2.rcr-stratified@0.1.keep20\endcsname{51.2}
\expandafter\def\csname ppt@rev.pol.keep.uiv.p2.rcr-stratified@0.1.keep20\endcsname{$+$51.2}
\expandafter\def\csname val@rev.pol.keep.uiv.p2.rcr-stratified@0.1.keepN\endcsname{0.582}
\expandafter\def\csname cnt@rev.pol.keep.uiv.p2.rcr-stratified@0.1.keepN\endcsname{1}
\expandafter\def\csname pct@rev.pol.keep.uiv.p2.rcr-stratified@0.1.keepN\endcsname{58.2}
\expandafter\def\csname ppt@rev.pol.keep.uiv.p2.rcr-stratified@0.1.keepN\endcsname{$+$58.2}
\expandafter\def\csname val@rev.pol.keep.uiv.p2.rcr-stratified@0.1.scoreN\endcsname{0.826}
\expandafter\def\csname cnt@rev.pol.keep.uiv.p2.rcr-stratified@0.1.scoreN\endcsname{1}
\expandafter\def\csname pct@rev.pol.keep.uiv.p2.rcr-stratified@0.1.scoreN\endcsname{82.6}
\expandafter\def\csname ppt@rev.pol.keep.uiv.p2.rcr-stratified@0.1.scoreN\endcsname{$+$82.6}
\expandafter\def\csname val@rev.pol.keep.uiv.p2.rcr-stratified@0.25.infeasible-rateN\endcsname{0.000}
\expandafter\def\csname cnt@rev.pol.keep.uiv.p2.rcr-stratified@0.25.infeasible-rateN\endcsname{0}
\expandafter\def\csname pct@rev.pol.keep.uiv.p2.rcr-stratified@0.25.infeasible-rateN\endcsname{0.0}
\expandafter\def\csname ppt@rev.pol.keep.uiv.p2.rcr-stratified@0.25.infeasible-rateN\endcsname{0.0}
\expandafter\def\csname val@rev.pol.keep.uiv.p2.rcr-stratified@0.25.keep20\endcsname{0.487}
\expandafter\def\csname cnt@rev.pol.keep.uiv.p2.rcr-stratified@0.25.keep20\endcsname{0}
\expandafter\def\csname pct@rev.pol.keep.uiv.p2.rcr-stratified@0.25.keep20\endcsname{48.8}
\expandafter\def\csname ppt@rev.pol.keep.uiv.p2.rcr-stratified@0.25.keep20\endcsname{$+$48.8}
\expandafter\def\csname val@rev.pol.keep.uiv.p2.rcr-stratified@0.25.keepN\endcsname{0.588}
\expandafter\def\csname cnt@rev.pol.keep.uiv.p2.rcr-stratified@0.25.keepN\endcsname{1}
\expandafter\def\csname pct@rev.pol.keep.uiv.p2.rcr-stratified@0.25.keepN\endcsname{58.8}
\expandafter\def\csname ppt@rev.pol.keep.uiv.p2.rcr-stratified@0.25.keepN\endcsname{$+$58.8}
\expandafter\def\csname val@rev.pol.keep.uiv.p2.rcr-stratified@0.25.scoreN\endcsname{0.828}
\expandafter\def\csname cnt@rev.pol.keep.uiv.p2.rcr-stratified@0.25.scoreN\endcsname{1}
\expandafter\def\csname pct@rev.pol.keep.uiv.p2.rcr-stratified@0.25.scoreN\endcsname{82.8}
\expandafter\def\csname ppt@rev.pol.keep.uiv.p2.rcr-stratified@0.25.scoreN\endcsname{$+$82.8}
\expandafter\def\csname val@rev.pol.keep.uiv.p2.rcr-stratified@0.5.infeasible-rateN\endcsname{0.000}
\expandafter\def\csname cnt@rev.pol.keep.uiv.p2.rcr-stratified@0.5.infeasible-rateN\endcsname{0}
\expandafter\def\csname pct@rev.pol.keep.uiv.p2.rcr-stratified@0.5.infeasible-rateN\endcsname{0.0}
\expandafter\def\csname ppt@rev.pol.keep.uiv.p2.rcr-stratified@0.5.infeasible-rateN\endcsname{0.0}
\expandafter\def\csname val@rev.pol.keep.uiv.p2.rcr-stratified@0.5.keep20\endcsname{0.425}
\expandafter\def\csname cnt@rev.pol.keep.uiv.p2.rcr-stratified@0.5.keep20\endcsname{0}
\expandafter\def\csname pct@rev.pol.keep.uiv.p2.rcr-stratified@0.5.keep20\endcsname{42.5}
\expandafter\def\csname ppt@rev.pol.keep.uiv.p2.rcr-stratified@0.5.keep20\endcsname{$+$42.5}
\expandafter\def\csname val@rev.pol.keep.uiv.p2.rcr-stratified@0.5.keepN\endcsname{0.603}
\expandafter\def\csname cnt@rev.pol.keep.uiv.p2.rcr-stratified@0.5.keepN\endcsname{1}
\expandafter\def\csname pct@rev.pol.keep.uiv.p2.rcr-stratified@0.5.keepN\endcsname{60.3}
\expandafter\def\csname ppt@rev.pol.keep.uiv.p2.rcr-stratified@0.5.keepN\endcsname{$+$60.3}
\expandafter\def\csname val@rev.pol.keep.uiv.p2.rcr-stratified@0.5.scoreN\endcsname{0.829}
\expandafter\def\csname cnt@rev.pol.keep.uiv.p2.rcr-stratified@0.5.scoreN\endcsname{1}
\expandafter\def\csname pct@rev.pol.keep.uiv.p2.rcr-stratified@0.5.scoreN\endcsname{82.9}
\expandafter\def\csname ppt@rev.pol.keep.uiv.p2.rcr-stratified@0.5.scoreN\endcsname{$+$82.9}
\expandafter\def\csname val@rev.pol.keep.uiv.p2.rcr-stratified@1.0.infeasible-rateN\endcsname{0.000}
\expandafter\def\csname cnt@rev.pol.keep.uiv.p2.rcr-stratified@1.0.infeasible-rateN\endcsname{0}
\expandafter\def\csname pct@rev.pol.keep.uiv.p2.rcr-stratified@1.0.infeasible-rateN\endcsname{0.0}
\expandafter\def\csname ppt@rev.pol.keep.uiv.p2.rcr-stratified@1.0.infeasible-rateN\endcsname{0.0}
\expandafter\def\csname val@rev.pol.keep.uiv.p2.rcr-stratified@1.0.keep20\endcsname{0.750}
\expandafter\def\csname cnt@rev.pol.keep.uiv.p2.rcr-stratified@1.0.keep20\endcsname{1}
\expandafter\def\csname pct@rev.pol.keep.uiv.p2.rcr-stratified@1.0.keep20\endcsname{75.0}
\expandafter\def\csname ppt@rev.pol.keep.uiv.p2.rcr-stratified@1.0.keep20\endcsname{$+$75.0}
\expandafter\def\csname val@rev.pol.keep.uiv.p2.rcr-stratified@1.0.keepN\endcsname{0.750}
\expandafter\def\csname cnt@rev.pol.keep.uiv.p2.rcr-stratified@1.0.keepN\endcsname{1}
\expandafter\def\csname pct@rev.pol.keep.uiv.p2.rcr-stratified@1.0.keepN\endcsname{75.0}
\expandafter\def\csname ppt@rev.pol.keep.uiv.p2.rcr-stratified@1.0.keepN\endcsname{$+$75.0}
\expandafter\def\csname val@rev.pol.keep.uiv.p2.rcr-stratified@1.0.scoreN\endcsname{0.832}
\expandafter\def\csname cnt@rev.pol.keep.uiv.p2.rcr-stratified@1.0.scoreN\endcsname{1}
\expandafter\def\csname pct@rev.pol.keep.uiv.p2.rcr-stratified@1.0.scoreN\endcsname{83.2}
\expandafter\def\csname ppt@rev.pol.keep.uiv.p2.rcr-stratified@1.0.scoreN\endcsname{$+$83.2}
\expandafter\def\csname val@rev.pol.keep.uiv.p2.replay-only.keep20\endcsname{0.500}
\expandafter\def\csname cnt@rev.pol.keep.uiv.p2.replay-only.keep20\endcsname{0}
\expandafter\def\csname pct@rev.pol.keep.uiv.p2.replay-only.keep20\endcsname{50.0}
\expandafter\def\csname ppt@rev.pol.keep.uiv.p2.replay-only.keep20\endcsname{$+$50.0}
\expandafter\def\csname val@rev.pol.keep.uiv.p2.replay-only.keepN\endcsname{0.500}
\expandafter\def\csname cnt@rev.pol.keep.uiv.p2.replay-only.keepN\endcsname{0}
\expandafter\def\csname pct@rev.pol.keep.uiv.p2.replay-only.keepN\endcsname{50.0}
\expandafter\def\csname ppt@rev.pol.keep.uiv.p2.replay-only.keepN\endcsname{$+$50.0}
\expandafter\def\csname val@rev.pol.keep.uiv.p2.replay-only.scoreN\endcsname{0.833}
\expandafter\def\csname cnt@rev.pol.keep.uiv.p2.replay-only.scoreN\endcsname{1}
\expandafter\def\csname pct@rev.pol.keep.uiv.p2.replay-only.scoreN\endcsname{83.3}
\expandafter\def\csname ppt@rev.pol.keep.uiv.p2.replay-only.scoreN\endcsname{$+$83.3}
\expandafter\def\csname val@rev.pol.keep.uiv.p2.replay-ppi-uniform@0.1.infeasible-rateN\endcsname{0.000}
\expandafter\def\csname cnt@rev.pol.keep.uiv.p2.replay-ppi-uniform@0.1.infeasible-rateN\endcsname{0}
\expandafter\def\csname pct@rev.pol.keep.uiv.p2.replay-ppi-uniform@0.1.infeasible-rateN\endcsname{0.0}
\expandafter\def\csname ppt@rev.pol.keep.uiv.p2.replay-ppi-uniform@0.1.infeasible-rateN\endcsname{0.0}
\expandafter\def\csname val@rev.pol.keep.uiv.p2.replay-ppi-uniform@0.1.keep20\endcsname{0.613}
\expandafter\def\csname cnt@rev.pol.keep.uiv.p2.replay-ppi-uniform@0.1.keep20\endcsname{1}
\expandafter\def\csname pct@rev.pol.keep.uiv.p2.replay-ppi-uniform@0.1.keep20\endcsname{61.3}
\expandafter\def\csname ppt@rev.pol.keep.uiv.p2.replay-ppi-uniform@0.1.keep20\endcsname{$+$61.3}
\expandafter\def\csname val@rev.pol.keep.uiv.p2.replay-ppi-uniform@0.1.keepN\endcsname{0.601}
\expandafter\def\csname cnt@rev.pol.keep.uiv.p2.replay-ppi-uniform@0.1.keepN\endcsname{1}
\expandafter\def\csname pct@rev.pol.keep.uiv.p2.replay-ppi-uniform@0.1.keepN\endcsname{60.1}
\expandafter\def\csname ppt@rev.pol.keep.uiv.p2.replay-ppi-uniform@0.1.keepN\endcsname{$+$60.1}
\expandafter\def\csname val@rev.pol.keep.uiv.p2.replay-ppi-uniform@0.1.scoreN\endcsname{0.829}
\expandafter\def\csname cnt@rev.pol.keep.uiv.p2.replay-ppi-uniform@0.1.scoreN\endcsname{1}
\expandafter\def\csname pct@rev.pol.keep.uiv.p2.replay-ppi-uniform@0.1.scoreN\endcsname{82.9}
\expandafter\def\csname ppt@rev.pol.keep.uiv.p2.replay-ppi-uniform@0.1.scoreN\endcsname{$+$82.9}
\expandafter\def\csname val@rev.pol.keep.uiv.p2.replay-ppi-uniform@0.25.infeasible-rateN\endcsname{0.000}
\expandafter\def\csname cnt@rev.pol.keep.uiv.p2.replay-ppi-uniform@0.25.infeasible-rateN\endcsname{0}
\expandafter\def\csname pct@rev.pol.keep.uiv.p2.replay-ppi-uniform@0.25.infeasible-rateN\endcsname{0.0}
\expandafter\def\csname ppt@rev.pol.keep.uiv.p2.replay-ppi-uniform@0.25.infeasible-rateN\endcsname{0.0}
\expandafter\def\csname val@rev.pol.keep.uiv.p2.replay-ppi-uniform@0.25.keep20\endcsname{0.500}
\expandafter\def\csname cnt@rev.pol.keep.uiv.p2.replay-ppi-uniform@0.25.keep20\endcsname{0}
\expandafter\def\csname pct@rev.pol.keep.uiv.p2.replay-ppi-uniform@0.25.keep20\endcsname{50.0}
\expandafter\def\csname ppt@rev.pol.keep.uiv.p2.replay-ppi-uniform@0.25.keep20\endcsname{$+$50.0}
\expandafter\def\csname val@rev.pol.keep.uiv.p2.replay-ppi-uniform@0.25.keepN\endcsname{0.580}
\expandafter\def\csname cnt@rev.pol.keep.uiv.p2.replay-ppi-uniform@0.25.keepN\endcsname{1}
\expandafter\def\csname pct@rev.pol.keep.uiv.p2.replay-ppi-uniform@0.25.keepN\endcsname{58.0}
\expandafter\def\csname ppt@rev.pol.keep.uiv.p2.replay-ppi-uniform@0.25.keepN\endcsname{$+$58.0}
\expandafter\def\csname val@rev.pol.keep.uiv.p2.replay-ppi-uniform@0.25.scoreN\endcsname{0.827}
\expandafter\def\csname cnt@rev.pol.keep.uiv.p2.replay-ppi-uniform@0.25.scoreN\endcsname{1}
\expandafter\def\csname pct@rev.pol.keep.uiv.p2.replay-ppi-uniform@0.25.scoreN\endcsname{82.7}
\expandafter\def\csname ppt@rev.pol.keep.uiv.p2.replay-ppi-uniform@0.25.scoreN\endcsname{$+$82.7}
\expandafter\def\csname val@rev.pol.keep.uiv.p2.replay-ppi-uniform@0.5.infeasible-rateN\endcsname{0.000}
\expandafter\def\csname cnt@rev.pol.keep.uiv.p2.replay-ppi-uniform@0.5.infeasible-rateN\endcsname{0}
\expandafter\def\csname pct@rev.pol.keep.uiv.p2.replay-ppi-uniform@0.5.infeasible-rateN\endcsname{0.0}
\expandafter\def\csname ppt@rev.pol.keep.uiv.p2.replay-ppi-uniform@0.5.infeasible-rateN\endcsname{0.0}
\expandafter\def\csname val@rev.pol.keep.uiv.p2.replay-ppi-uniform@0.5.keep20\endcsname{0.675}
\expandafter\def\csname cnt@rev.pol.keep.uiv.p2.replay-ppi-uniform@0.5.keep20\endcsname{1}
\expandafter\def\csname pct@rev.pol.keep.uiv.p2.replay-ppi-uniform@0.5.keep20\endcsname{67.5}
\expandafter\def\csname ppt@rev.pol.keep.uiv.p2.replay-ppi-uniform@0.5.keep20\endcsname{$+$67.5}
\expandafter\def\csname val@rev.pol.keep.uiv.p2.replay-ppi-uniform@0.5.keepN\endcsname{0.688}
\expandafter\def\csname cnt@rev.pol.keep.uiv.p2.replay-ppi-uniform@0.5.keepN\endcsname{1}
\expandafter\def\csname pct@rev.pol.keep.uiv.p2.replay-ppi-uniform@0.5.keepN\endcsname{68.8}
\expandafter\def\csname ppt@rev.pol.keep.uiv.p2.replay-ppi-uniform@0.5.keepN\endcsname{$+$68.8}
\expandafter\def\csname val@rev.pol.keep.uiv.p2.replay-ppi-uniform@0.5.scoreN\endcsname{0.826}
\expandafter\def\csname cnt@rev.pol.keep.uiv.p2.replay-ppi-uniform@0.5.scoreN\endcsname{1}
\expandafter\def\csname pct@rev.pol.keep.uiv.p2.replay-ppi-uniform@0.5.scoreN\endcsname{82.6}
\expandafter\def\csname ppt@rev.pol.keep.uiv.p2.replay-ppi-uniform@0.5.scoreN\endcsname{$+$82.6}
\expandafter\def\csname val@rev.pol.keep.uiv.p2.replay-ppi-uniform@1.0.infeasible-rateN\endcsname{0.000}
\expandafter\def\csname cnt@rev.pol.keep.uiv.p2.replay-ppi-uniform@1.0.infeasible-rateN\endcsname{0}
\expandafter\def\csname pct@rev.pol.keep.uiv.p2.replay-ppi-uniform@1.0.infeasible-rateN\endcsname{0.0}
\expandafter\def\csname ppt@rev.pol.keep.uiv.p2.replay-ppi-uniform@1.0.infeasible-rateN\endcsname{0.0}
\expandafter\def\csname val@rev.pol.keep.uiv.p2.replay-ppi-uniform@1.0.keep20\endcsname{0.750}
\expandafter\def\csname cnt@rev.pol.keep.uiv.p2.replay-ppi-uniform@1.0.keep20\endcsname{1}
\expandafter\def\csname pct@rev.pol.keep.uiv.p2.replay-ppi-uniform@1.0.keep20\endcsname{75.0}
\expandafter\def\csname ppt@rev.pol.keep.uiv.p2.replay-ppi-uniform@1.0.keep20\endcsname{$+$75.0}
\expandafter\def\csname val@rev.pol.keep.uiv.p2.replay-ppi-uniform@1.0.keepN\endcsname{0.750}
\expandafter\def\csname cnt@rev.pol.keep.uiv.p2.replay-ppi-uniform@1.0.keepN\endcsname{1}
\expandafter\def\csname pct@rev.pol.keep.uiv.p2.replay-ppi-uniform@1.0.keepN\endcsname{75.0}
\expandafter\def\csname ppt@rev.pol.keep.uiv.p2.replay-ppi-uniform@1.0.keepN\endcsname{$+$75.0}
\expandafter\def\csname val@rev.pol.keep.uiv.p2.replay-ppi-uniform@1.0.scoreN\endcsname{0.832}
\expandafter\def\csname cnt@rev.pol.keep.uiv.p2.replay-ppi-uniform@1.0.scoreN\endcsname{1}
\expandafter\def\csname pct@rev.pol.keep.uiv.p2.replay-ppi-uniform@1.0.scoreN\endcsname{83.2}
\expandafter\def\csname ppt@rev.pol.keep.uiv.p2.replay-ppi-uniform@1.0.scoreN\endcsname{$+$83.2}
\expandafter\def\csname val@rev.pol.keep.uiv.p2.suspend-all.keep20\endcsname{0.000}
\expandafter\def\csname cnt@rev.pol.keep.uiv.p2.suspend-all.keep20\endcsname{0}
\expandafter\def\csname pct@rev.pol.keep.uiv.p2.suspend-all.keep20\endcsname{0.0}
\expandafter\def\csname ppt@rev.pol.keep.uiv.p2.suspend-all.keep20\endcsname{0.0}
\expandafter\def\csname val@rev.pol.keep.uiv.p2.suspend-all.keepN\endcsname{0.000}
\expandafter\def\csname cnt@rev.pol.keep.uiv.p2.suspend-all.keepN\endcsname{0}
\expandafter\def\csname pct@rev.pol.keep.uiv.p2.suspend-all.keepN\endcsname{0.0}
\expandafter\def\csname ppt@rev.pol.keep.uiv.p2.suspend-all.keepN\endcsname{0.0}
\expandafter\def\csname val@rev.pol.keep.uiv.p2.suspend-all.scoreN\endcsname{0.840}
\expandafter\def\csname cnt@rev.pol.keep.uiv.p2.suspend-all.scoreN\endcsname{1}
\expandafter\def\csname pct@rev.pol.keep.uiv.p2.suspend-all.scoreN\endcsname{84.0}
\expandafter\def\csname ppt@rev.pol.keep.uiv.p2.suspend-all.scoreN\endcsname{$+$84.0}
\expandafter\def\csname val@rev.pol.keep.uiv.p2.uniform-fresh@0.1.infeasible-rateN\endcsname{0.000}
\expandafter\def\csname cnt@rev.pol.keep.uiv.p2.uniform-fresh@0.1.infeasible-rateN\endcsname{0}
\expandafter\def\csname pct@rev.pol.keep.uiv.p2.uniform-fresh@0.1.infeasible-rateN\endcsname{0.0}
\expandafter\def\csname ppt@rev.pol.keep.uiv.p2.uniform-fresh@0.1.infeasible-rateN\endcsname{0.0}
\expandafter\def\csname val@rev.pol.keep.uiv.p2.uniform-fresh@0.1.keep20\endcsname{0.838}
\expandafter\def\csname cnt@rev.pol.keep.uiv.p2.uniform-fresh@0.1.keep20\endcsname{1}
\expandafter\def\csname pct@rev.pol.keep.uiv.p2.uniform-fresh@0.1.keep20\endcsname{83.8}
\expandafter\def\csname ppt@rev.pol.keep.uiv.p2.uniform-fresh@0.1.keep20\endcsname{$+$83.8}
\expandafter\def\csname val@rev.pol.keep.uiv.p2.uniform-fresh@0.1.keepN\endcsname{0.778}
\expandafter\def\csname cnt@rev.pol.keep.uiv.p2.uniform-fresh@0.1.keepN\endcsname{1}
\expandafter\def\csname pct@rev.pol.keep.uiv.p2.uniform-fresh@0.1.keepN\endcsname{77.8}
\expandafter\def\csname ppt@rev.pol.keep.uiv.p2.uniform-fresh@0.1.keepN\endcsname{$+$77.8}
\expandafter\def\csname val@rev.pol.keep.uiv.p2.uniform-fresh@0.1.scoreN\endcsname{0.822}
\expandafter\def\csname cnt@rev.pol.keep.uiv.p2.uniform-fresh@0.1.scoreN\endcsname{1}
\expandafter\def\csname pct@rev.pol.keep.uiv.p2.uniform-fresh@0.1.scoreN\endcsname{82.2}
\expandafter\def\csname ppt@rev.pol.keep.uiv.p2.uniform-fresh@0.1.scoreN\endcsname{$+$82.2}
\expandafter\def\csname val@rev.pol.keep.uiv.p2.uniform-fresh@0.25.infeasible-rateN\endcsname{0.000}
\expandafter\def\csname cnt@rev.pol.keep.uiv.p2.uniform-fresh@0.25.infeasible-rateN\endcsname{0}
\expandafter\def\csname pct@rev.pol.keep.uiv.p2.uniform-fresh@0.25.infeasible-rateN\endcsname{0.0}
\expandafter\def\csname ppt@rev.pol.keep.uiv.p2.uniform-fresh@0.25.infeasible-rateN\endcsname{0.0}
\expandafter\def\csname val@rev.pol.keep.uiv.p2.uniform-fresh@0.25.keep20\endcsname{0.738}
\expandafter\def\csname cnt@rev.pol.keep.uiv.p2.uniform-fresh@0.25.keep20\endcsname{1}
\expandafter\def\csname pct@rev.pol.keep.uiv.p2.uniform-fresh@0.25.keep20\endcsname{73.8}
\expandafter\def\csname ppt@rev.pol.keep.uiv.p2.uniform-fresh@0.25.keep20\endcsname{$+$73.8}
\expandafter\def\csname val@rev.pol.keep.uiv.p2.uniform-fresh@0.25.keepN\endcsname{0.718}
\expandafter\def\csname cnt@rev.pol.keep.uiv.p2.uniform-fresh@0.25.keepN\endcsname{1}
\expandafter\def\csname pct@rev.pol.keep.uiv.p2.uniform-fresh@0.25.keepN\endcsname{71.8}
\expandafter\def\csname ppt@rev.pol.keep.uiv.p2.uniform-fresh@0.25.keepN\endcsname{$+$71.8}
\expandafter\def\csname val@rev.pol.keep.uiv.p2.uniform-fresh@0.25.scoreN\endcsname{0.824}
\expandafter\def\csname cnt@rev.pol.keep.uiv.p2.uniform-fresh@0.25.scoreN\endcsname{1}
\expandafter\def\csname pct@rev.pol.keep.uiv.p2.uniform-fresh@0.25.scoreN\endcsname{82.4}
\expandafter\def\csname ppt@rev.pol.keep.uiv.p2.uniform-fresh@0.25.scoreN\endcsname{$+$82.4}
\expandafter\def\csname val@rev.pol.keep.uiv.p2.uniform-fresh@0.5.infeasible-rateN\endcsname{0.000}
\expandafter\def\csname cnt@rev.pol.keep.uiv.p2.uniform-fresh@0.5.infeasible-rateN\endcsname{0}
\expandafter\def\csname pct@rev.pol.keep.uiv.p2.uniform-fresh@0.5.infeasible-rateN\endcsname{0.0}
\expandafter\def\csname ppt@rev.pol.keep.uiv.p2.uniform-fresh@0.5.infeasible-rateN\endcsname{0.0}
\expandafter\def\csname val@rev.pol.keep.uiv.p2.uniform-fresh@0.5.keep20\endcsname{0.650}
\expandafter\def\csname cnt@rev.pol.keep.uiv.p2.uniform-fresh@0.5.keep20\endcsname{1}
\expandafter\def\csname pct@rev.pol.keep.uiv.p2.uniform-fresh@0.5.keep20\endcsname{65.0}
\expandafter\def\csname ppt@rev.pol.keep.uiv.p2.uniform-fresh@0.5.keep20\endcsname{$+$65.0}
\expandafter\def\csname val@rev.pol.keep.uiv.p2.uniform-fresh@0.5.keepN\endcsname{0.726}
\expandafter\def\csname cnt@rev.pol.keep.uiv.p2.uniform-fresh@0.5.keepN\endcsname{1}
\expandafter\def\csname pct@rev.pol.keep.uiv.p2.uniform-fresh@0.5.keepN\endcsname{72.6}
\expandafter\def\csname ppt@rev.pol.keep.uiv.p2.uniform-fresh@0.5.keepN\endcsname{$+$72.6}
\expandafter\def\csname val@rev.pol.keep.uiv.p2.uniform-fresh@0.5.scoreN\endcsname{0.825}
\expandafter\def\csname cnt@rev.pol.keep.uiv.p2.uniform-fresh@0.5.scoreN\endcsname{1}
\expandafter\def\csname pct@rev.pol.keep.uiv.p2.uniform-fresh@0.5.scoreN\endcsname{82.5}
\expandafter\def\csname ppt@rev.pol.keep.uiv.p2.uniform-fresh@0.5.scoreN\endcsname{$+$82.5}
\expandafter\def\csname val@rev.pol.keep.uiv.p2.uniform-fresh@1.0.infeasible-rateN\endcsname{0.000}
\expandafter\def\csname cnt@rev.pol.keep.uiv.p2.uniform-fresh@1.0.infeasible-rateN\endcsname{0}
\expandafter\def\csname pct@rev.pol.keep.uiv.p2.uniform-fresh@1.0.infeasible-rateN\endcsname{0.0}
\expandafter\def\csname ppt@rev.pol.keep.uiv.p2.uniform-fresh@1.0.infeasible-rateN\endcsname{0.0}
\expandafter\def\csname val@rev.pol.keep.uiv.p2.uniform-fresh@1.0.keep20\endcsname{0.750}
\expandafter\def\csname cnt@rev.pol.keep.uiv.p2.uniform-fresh@1.0.keep20\endcsname{1}
\expandafter\def\csname pct@rev.pol.keep.uiv.p2.uniform-fresh@1.0.keep20\endcsname{75.0}
\expandafter\def\csname ppt@rev.pol.keep.uiv.p2.uniform-fresh@1.0.keep20\endcsname{$+$75.0}
\expandafter\def\csname val@rev.pol.keep.uiv.p2.uniform-fresh@1.0.keepN\endcsname{0.750}
\expandafter\def\csname cnt@rev.pol.keep.uiv.p2.uniform-fresh@1.0.keepN\endcsname{1}
\expandafter\def\csname pct@rev.pol.keep.uiv.p2.uniform-fresh@1.0.keepN\endcsname{75.0}
\expandafter\def\csname ppt@rev.pol.keep.uiv.p2.uniform-fresh@1.0.keepN\endcsname{$+$75.0}
\expandafter\def\csname val@rev.pol.keep.uiv.p2.uniform-fresh@1.0.scoreN\endcsname{0.832}
\expandafter\def\csname cnt@rev.pol.keep.uiv.p2.uniform-fresh@1.0.scoreN\endcsname{1}
\expandafter\def\csname pct@rev.pol.keep.uiv.p2.uniform-fresh@1.0.scoreN\endcsname{83.2}
\expandafter\def\csname ppt@rev.pol.keep.uiv.p2.uniform-fresh@1.0.scoreN\endcsname{$+$83.2}
\expandafter\def\csname val@rev.pol.mdd-tau.sspro.max\endcsname{0.209}
\expandafter\def\csname cnt@rev.pol.mdd-tau.sspro.max\endcsname{0}
\expandafter\def\csname pct@rev.pol.mdd-tau.sspro.max\endcsname{20.9}
\expandafter\def\csname ppt@rev.pol.mdd-tau.sspro.max\endcsname{$+$20.9}
\expandafter\def\csname val@rev.pol.mdd-tau.sspro.min\endcsname{0.102}
\expandafter\def\csname cnt@rev.pol.mdd-tau.sspro.min\endcsname{0}
\expandafter\def\csname pct@rev.pol.mdd-tau.sspro.min\endcsname{10.2}
\expandafter\def\csname ppt@rev.pol.mdd-tau.sspro.min\endcsname{$+$10.2}
\expandafter\def\csname val@rev.pol.mdd-tau.sspro.n\endcsname{8}
\expandafter\def\csname cnt@rev.pol.mdd-tau.sspro.n\endcsname{8}
\expandafter\def\csname pct@rev.pol.mdd-tau.sspro.n\endcsname{800.0}
\expandafter\def\csname ppt@rev.pol.mdd-tau.sspro.n\endcsname{$+$800.0}
\expandafter\def\csname val@rev.pol.mdd-tau.sspro.n-detectable\endcsname{4}
\expandafter\def\csname cnt@rev.pol.mdd-tau.sspro.n-detectable\endcsname{4}
\expandafter\def\csname pct@rev.pol.mdd-tau.sspro.n-detectable\endcsname{400.0}
\expandafter\def\csname ppt@rev.pol.mdd-tau.sspro.n-detectable\endcsname{$+$400.0}
\expandafter\def\csname val@rev.pol.mdd-tau.sspro.p1.center-d1.G\endcsname{12}
\expandafter\def\csname cnt@rev.pol.mdd-tau.sspro.p1.center-d1.G\endcsname{12}
\expandafter\def\csname pct@rev.pol.mdd-tau.sspro.p1.center-d1.G\endcsname{1200.0}
\expandafter\def\csname ppt@rev.pol.mdd-tau.sspro.p1.center-d1.G\endcsname{$+$1200.0}
\expandafter\def\csname val@rev.pol.mdd-tau.sspro.p1.center-d1.detectable\endcsname{0}
\expandafter\def\csname cnt@rev.pol.mdd-tau.sspro.p1.center-d1.detectable\endcsname{0}
\expandafter\def\csname pct@rev.pol.mdd-tau.sspro.p1.center-d1.detectable\endcsname{0.0}
\expandafter\def\csname ppt@rev.pol.mdd-tau.sspro.p1.center-d1.detectable\endcsname{0.0}
\expandafter\def\csname val@rev.pol.mdd-tau.sspro.p1.center-d1.mdd\endcsname{0.147}
\expandafter\def\csname cnt@rev.pol.mdd-tau.sspro.p1.center-d1.mdd\endcsname{0}
\expandafter\def\csname pct@rev.pol.mdd-tau.sspro.p1.center-d1.mdd\endcsname{14.7}
\expandafter\def\csname ppt@rev.pol.mdd-tau.sspro.p1.center-d1.mdd\endcsname{$+$14.7}
\expandafter\def\csname val@rev.pol.mdd-tau.sspro.p1.center-d1.obs\endcsname{0.104}
\expandafter\def\csname cnt@rev.pol.mdd-tau.sspro.p1.center-d1.obs\endcsname{0}
\expandafter\def\csname pct@rev.pol.mdd-tau.sspro.p1.center-d1.obs\endcsname{10.4}
\expandafter\def\csname ppt@rev.pol.mdd-tau.sspro.p1.center-d1.obs\endcsname{$+$10.4}
\expandafter\def\csname val@rev.pol.mdd-tau.sspro.p1.center-d1.sd\endcsname{0.191}
\expandafter\def\csname cnt@rev.pol.mdd-tau.sspro.p1.center-d1.sd\endcsname{0}
\expandafter\def\csname pct@rev.pol.mdd-tau.sspro.p1.center-d1.sd\endcsname{19.1}
\expandafter\def\csname ppt@rev.pol.mdd-tau.sspro.p1.center-d1.sd\endcsname{$+$19.1}
\expandafter\def\csname val@rev.pol.mdd-tau.sspro.p1.context-d1.G\endcsname{12}
\expandafter\def\csname cnt@rev.pol.mdd-tau.sspro.p1.context-d1.G\endcsname{12}
\expandafter\def\csname pct@rev.pol.mdd-tau.sspro.p1.context-d1.G\endcsname{1200.0}
\expandafter\def\csname ppt@rev.pol.mdd-tau.sspro.p1.context-d1.G\endcsname{$+$1200.0}
\expandafter\def\csname val@rev.pol.mdd-tau.sspro.p1.context-d1.detectable\endcsname{0}
\expandafter\def\csname cnt@rev.pol.mdd-tau.sspro.p1.context-d1.detectable\endcsname{0}
\expandafter\def\csname pct@rev.pol.mdd-tau.sspro.p1.context-d1.detectable\endcsname{0.0}
\expandafter\def\csname ppt@rev.pol.mdd-tau.sspro.p1.context-d1.detectable\endcsname{0.0}
\expandafter\def\csname val@rev.pol.mdd-tau.sspro.p1.context-d1.mdd\endcsname{0.172}
\expandafter\def\csname cnt@rev.pol.mdd-tau.sspro.p1.context-d1.mdd\endcsname{0}
\expandafter\def\csname pct@rev.pol.mdd-tau.sspro.p1.context-d1.mdd\endcsname{17.2}
\expandafter\def\csname ppt@rev.pol.mdd-tau.sspro.p1.context-d1.mdd\endcsname{$+$17.2}
\expandafter\def\csname val@rev.pol.mdd-tau.sspro.p1.context-d1.obs\endcsname{0.115}
\expandafter\def\csname cnt@rev.pol.mdd-tau.sspro.p1.context-d1.obs\endcsname{0}
\expandafter\def\csname pct@rev.pol.mdd-tau.sspro.p1.context-d1.obs\endcsname{11.5}
\expandafter\def\csname ppt@rev.pol.mdd-tau.sspro.p1.context-d1.obs\endcsname{$+$11.5}
\expandafter\def\csname val@rev.pol.mdd-tau.sspro.p1.context-d1.sd\endcsname{0.223}
\expandafter\def\csname cnt@rev.pol.mdd-tau.sspro.p1.context-d1.sd\endcsname{0}
\expandafter\def\csname pct@rev.pol.mdd-tau.sspro.p1.context-d1.sd\endcsname{22.3}
\expandafter\def\csname ppt@rev.pol.mdd-tau.sspro.p1.context-d1.sd\endcsname{$+$22.3}
\expandafter\def\csname val@rev.pol.mdd-tau.sspro.p1.dual-d2.G\endcsname{12}
\expandafter\def\csname cnt@rev.pol.mdd-tau.sspro.p1.dual-d2.G\endcsname{12}
\expandafter\def\csname pct@rev.pol.mdd-tau.sspro.p1.dual-d2.G\endcsname{1200.0}
\expandafter\def\csname ppt@rev.pol.mdd-tau.sspro.p1.dual-d2.G\endcsname{$+$1200.0}
\expandafter\def\csname val@rev.pol.mdd-tau.sspro.p1.dual-d2.detectable\endcsname{0}
\expandafter\def\csname cnt@rev.pol.mdd-tau.sspro.p1.dual-d2.detectable\endcsname{0}
\expandafter\def\csname pct@rev.pol.mdd-tau.sspro.p1.dual-d2.detectable\endcsname{0.0}
\expandafter\def\csname ppt@rev.pol.mdd-tau.sspro.p1.dual-d2.detectable\endcsname{0.0}
\expandafter\def\csname val@rev.pol.mdd-tau.sspro.p1.dual-d2.mdd\endcsname{0.209}
\expandafter\def\csname cnt@rev.pol.mdd-tau.sspro.p1.dual-d2.mdd\endcsname{0}
\expandafter\def\csname pct@rev.pol.mdd-tau.sspro.p1.dual-d2.mdd\endcsname{20.9}
\expandafter\def\csname ppt@rev.pol.mdd-tau.sspro.p1.dual-d2.mdd\endcsname{$+$20.9}
\expandafter\def\csname val@rev.pol.mdd-tau.sspro.p1.dual-d2.obs\endcsname{0.146}
\expandafter\def\csname cnt@rev.pol.mdd-tau.sspro.p1.dual-d2.obs\endcsname{0}
\expandafter\def\csname pct@rev.pol.mdd-tau.sspro.p1.dual-d2.obs\endcsname{14.6}
\expandafter\def\csname ppt@rev.pol.mdd-tau.sspro.p1.dual-d2.obs\endcsname{$+$14.6}
\expandafter\def\csname val@rev.pol.mdd-tau.sspro.p1.dual-d2.sd\endcsname{0.271}
\expandafter\def\csname cnt@rev.pol.mdd-tau.sspro.p1.dual-d2.sd\endcsname{0}
\expandafter\def\csname pct@rev.pol.mdd-tau.sspro.p1.dual-d2.sd\endcsname{27.1}
\expandafter\def\csname ppt@rev.pol.mdd-tau.sspro.p1.dual-d2.sd\endcsname{$+$27.1}
\expandafter\def\csname val@rev.pol.mdd-tau.sspro.p1.grid2-d3.G\endcsname{12}
\expandafter\def\csname cnt@rev.pol.mdd-tau.sspro.p1.grid2-d3.G\endcsname{12}
\expandafter\def\csname pct@rev.pol.mdd-tau.sspro.p1.grid2-d3.G\endcsname{1200.0}
\expandafter\def\csname ppt@rev.pol.mdd-tau.sspro.p1.grid2-d3.G\endcsname{$+$1200.0}
\expandafter\def\csname val@rev.pol.mdd-tau.sspro.p1.grid2-d3.detectable\endcsname{0}
\expandafter\def\csname cnt@rev.pol.mdd-tau.sspro.p1.grid2-d3.detectable\endcsname{0}
\expandafter\def\csname pct@rev.pol.mdd-tau.sspro.p1.grid2-d3.detectable\endcsname{0.0}
\expandafter\def\csname ppt@rev.pol.mdd-tau.sspro.p1.grid2-d3.detectable\endcsname{0.0}
\expandafter\def\csname val@rev.pol.mdd-tau.sspro.p1.grid2-d3.mdd\endcsname{0.167}
\expandafter\def\csname cnt@rev.pol.mdd-tau.sspro.p1.grid2-d3.mdd\endcsname{0}
\expandafter\def\csname pct@rev.pol.mdd-tau.sspro.p1.grid2-d3.mdd\endcsname{16.7}
\expandafter\def\csname ppt@rev.pol.mdd-tau.sspro.p1.grid2-d3.mdd\endcsname{$+$16.7}
\expandafter\def\csname val@rev.pol.mdd-tau.sspro.p1.grid2-d3.obs\endcsname{0.115}
\expandafter\def\csname cnt@rev.pol.mdd-tau.sspro.p1.grid2-d3.obs\endcsname{0}
\expandafter\def\csname pct@rev.pol.mdd-tau.sspro.p1.grid2-d3.obs\endcsname{11.5}
\expandafter\def\csname ppt@rev.pol.mdd-tau.sspro.p1.grid2-d3.obs\endcsname{$+$11.5}
\expandafter\def\csname val@rev.pol.mdd-tau.sspro.p1.grid2-d3.sd\endcsname{0.216}
\expandafter\def\csname cnt@rev.pol.mdd-tau.sspro.p1.grid2-d3.sd\endcsname{0}
\expandafter\def\csname pct@rev.pol.mdd-tau.sspro.p1.grid2-d3.sd\endcsname{21.6}
\expandafter\def\csname ppt@rev.pol.mdd-tau.sspro.p1.grid2-d3.sd\endcsname{$+$21.6}
\expandafter\def\csname val@rev.pol.mdd-tau.sspro.p1.zoomclick-d3.G\endcsname{12}
\expandafter\def\csname cnt@rev.pol.mdd-tau.sspro.p1.zoomclick-d3.G\endcsname{12}
\expandafter\def\csname pct@rev.pol.mdd-tau.sspro.p1.zoomclick-d3.G\endcsname{1200.0}
\expandafter\def\csname ppt@rev.pol.mdd-tau.sspro.p1.zoomclick-d3.G\endcsname{$+$1200.0}
\expandafter\def\csname val@rev.pol.mdd-tau.sspro.p1.zoomclick-d3.detectable\endcsname{1}
\expandafter\def\csname cnt@rev.pol.mdd-tau.sspro.p1.zoomclick-d3.detectable\endcsname{1}
\expandafter\def\csname pct@rev.pol.mdd-tau.sspro.p1.zoomclick-d3.detectable\endcsname{100.0}
\expandafter\def\csname ppt@rev.pol.mdd-tau.sspro.p1.zoomclick-d3.detectable\endcsname{$+$100.0}
\expandafter\def\csname val@rev.pol.mdd-tau.sspro.p1.zoomclick-d3.mdd\endcsname{0.139}
\expandafter\def\csname cnt@rev.pol.mdd-tau.sspro.p1.zoomclick-d3.mdd\endcsname{0}
\expandafter\def\csname pct@rev.pol.mdd-tau.sspro.p1.zoomclick-d3.mdd\endcsname{13.9}
\expandafter\def\csname ppt@rev.pol.mdd-tau.sspro.p1.zoomclick-d3.mdd\endcsname{$+$13.9}
\expandafter\def\csname val@rev.pol.mdd-tau.sspro.p1.zoomclick-d3.obs\endcsname{0.177}
\expandafter\def\csname cnt@rev.pol.mdd-tau.sspro.p1.zoomclick-d3.obs\endcsname{0}
\expandafter\def\csname pct@rev.pol.mdd-tau.sspro.p1.zoomclick-d3.obs\endcsname{17.7}
\expandafter\def\csname ppt@rev.pol.mdd-tau.sspro.p1.zoomclick-d3.obs\endcsname{$+$17.7}
\expandafter\def\csname val@rev.pol.mdd-tau.sspro.p1.zoomclick-d3.sd\endcsname{0.180}
\expandafter\def\csname cnt@rev.pol.mdd-tau.sspro.p1.zoomclick-d3.sd\endcsname{0}
\expandafter\def\csname pct@rev.pol.mdd-tau.sspro.p1.zoomclick-d3.sd\endcsname{18.0}
\expandafter\def\csname ppt@rev.pol.mdd-tau.sspro.p1.zoomclick-d3.sd\endcsname{$+$18.0}
\expandafter\def\csname val@rev.pol.mdd-tau.sspro.p2.center-d1.G\endcsname{12}
\expandafter\def\csname cnt@rev.pol.mdd-tau.sspro.p2.center-d1.G\endcsname{12}
\expandafter\def\csname pct@rev.pol.mdd-tau.sspro.p2.center-d1.G\endcsname{1200.0}
\expandafter\def\csname ppt@rev.pol.mdd-tau.sspro.p2.center-d1.G\endcsname{$+$1200.0}
\expandafter\def\csname val@rev.pol.mdd-tau.sspro.p2.center-d1.detectable\endcsname{1}
\expandafter\def\csname cnt@rev.pol.mdd-tau.sspro.p2.center-d1.detectable\endcsname{1}
\expandafter\def\csname pct@rev.pol.mdd-tau.sspro.p2.center-d1.detectable\endcsname{100.0}
\expandafter\def\csname ppt@rev.pol.mdd-tau.sspro.p2.center-d1.detectable\endcsname{$+$100.0}
\expandafter\def\csname val@rev.pol.mdd-tau.sspro.p2.center-d1.mdd\endcsname{0.110}
\expandafter\def\csname cnt@rev.pol.mdd-tau.sspro.p2.center-d1.mdd\endcsname{0}
\expandafter\def\csname pct@rev.pol.mdd-tau.sspro.p2.center-d1.mdd\endcsname{11.0}
\expandafter\def\csname ppt@rev.pol.mdd-tau.sspro.p2.center-d1.mdd\endcsname{$+$11.0}
\expandafter\def\csname val@rev.pol.mdd-tau.sspro.p2.center-d1.obs\endcsname{0.156}
\expandafter\def\csname cnt@rev.pol.mdd-tau.sspro.p2.center-d1.obs\endcsname{0}
\expandafter\def\csname pct@rev.pol.mdd-tau.sspro.p2.center-d1.obs\endcsname{15.6}
\expandafter\def\csname ppt@rev.pol.mdd-tau.sspro.p2.center-d1.obs\endcsname{$+$15.6}
\expandafter\def\csname val@rev.pol.mdd-tau.sspro.p2.center-d1.sd\endcsname{0.142}
\expandafter\def\csname cnt@rev.pol.mdd-tau.sspro.p2.center-d1.sd\endcsname{0}
\expandafter\def\csname pct@rev.pol.mdd-tau.sspro.p2.center-d1.sd\endcsname{14.2}
\expandafter\def\csname ppt@rev.pol.mdd-tau.sspro.p2.center-d1.sd\endcsname{$+$14.2}
\expandafter\def\csname val@rev.pol.mdd-tau.sspro.p2.context-d1.G\endcsname{12}
\expandafter\def\csname cnt@rev.pol.mdd-tau.sspro.p2.context-d1.G\endcsname{12}
\expandafter\def\csname pct@rev.pol.mdd-tau.sspro.p2.context-d1.G\endcsname{1200.0}
\expandafter\def\csname ppt@rev.pol.mdd-tau.sspro.p2.context-d1.G\endcsname{$+$1200.0}
\expandafter\def\csname val@rev.pol.mdd-tau.sspro.p2.context-d1.detectable\endcsname{1}
\expandafter\def\csname cnt@rev.pol.mdd-tau.sspro.p2.context-d1.detectable\endcsname{1}
\expandafter\def\csname pct@rev.pol.mdd-tau.sspro.p2.context-d1.detectable\endcsname{100.0}
\expandafter\def\csname ppt@rev.pol.mdd-tau.sspro.p2.context-d1.detectable\endcsname{$+$100.0}
\expandafter\def\csname val@rev.pol.mdd-tau.sspro.p2.context-d1.mdd\endcsname{0.109}
\expandafter\def\csname cnt@rev.pol.mdd-tau.sspro.p2.context-d1.mdd\endcsname{0}
\expandafter\def\csname pct@rev.pol.mdd-tau.sspro.p2.context-d1.mdd\endcsname{10.9}
\expandafter\def\csname ppt@rev.pol.mdd-tau.sspro.p2.context-d1.mdd\endcsname{$+$10.9}
\expandafter\def\csname val@rev.pol.mdd-tau.sspro.p2.context-d1.obs\endcsname{0.125}
\expandafter\def\csname cnt@rev.pol.mdd-tau.sspro.p2.context-d1.obs\endcsname{0}
\expandafter\def\csname pct@rev.pol.mdd-tau.sspro.p2.context-d1.obs\endcsname{12.5}
\expandafter\def\csname ppt@rev.pol.mdd-tau.sspro.p2.context-d1.obs\endcsname{$+$12.5}
\expandafter\def\csname val@rev.pol.mdd-tau.sspro.p2.context-d1.sd\endcsname{0.141}
\expandafter\def\csname cnt@rev.pol.mdd-tau.sspro.p2.context-d1.sd\endcsname{0}
\expandafter\def\csname pct@rev.pol.mdd-tau.sspro.p2.context-d1.sd\endcsname{14.1}
\expandafter\def\csname ppt@rev.pol.mdd-tau.sspro.p2.context-d1.sd\endcsname{$+$14.1}
\expandafter\def\csname val@rev.pol.mdd-tau.sspro.p2.dual-d2.G\endcsname{12}
\expandafter\def\csname cnt@rev.pol.mdd-tau.sspro.p2.dual-d2.G\endcsname{12}
\expandafter\def\csname pct@rev.pol.mdd-tau.sspro.p2.dual-d2.G\endcsname{1200.0}
\expandafter\def\csname ppt@rev.pol.mdd-tau.sspro.p2.dual-d2.G\endcsname{$+$1200.0}
\expandafter\def\csname val@rev.pol.mdd-tau.sspro.p2.dual-d2.detectable\endcsname{1}
\expandafter\def\csname cnt@rev.pol.mdd-tau.sspro.p2.dual-d2.detectable\endcsname{1}
\expandafter\def\csname pct@rev.pol.mdd-tau.sspro.p2.dual-d2.detectable\endcsname{100.0}
\expandafter\def\csname ppt@rev.pol.mdd-tau.sspro.p2.dual-d2.detectable\endcsname{$+$100.0}
\expandafter\def\csname val@rev.pol.mdd-tau.sspro.p2.dual-d2.mdd\endcsname{0.102}
\expandafter\def\csname cnt@rev.pol.mdd-tau.sspro.p2.dual-d2.mdd\endcsname{0}
\expandafter\def\csname pct@rev.pol.mdd-tau.sspro.p2.dual-d2.mdd\endcsname{10.2}
\expandafter\def\csname ppt@rev.pol.mdd-tau.sspro.p2.dual-d2.mdd\endcsname{$+$10.2}
\expandafter\def\csname val@rev.pol.mdd-tau.sspro.p2.dual-d2.obs\endcsname{0.156}
\expandafter\def\csname cnt@rev.pol.mdd-tau.sspro.p2.dual-d2.obs\endcsname{0}
\expandafter\def\csname pct@rev.pol.mdd-tau.sspro.p2.dual-d2.obs\endcsname{15.6}
\expandafter\def\csname ppt@rev.pol.mdd-tau.sspro.p2.dual-d2.obs\endcsname{$+$15.6}
\expandafter\def\csname val@rev.pol.mdd-tau.sspro.p2.dual-d2.sd\endcsname{0.132}
\expandafter\def\csname cnt@rev.pol.mdd-tau.sspro.p2.dual-d2.sd\endcsname{0}
\expandafter\def\csname pct@rev.pol.mdd-tau.sspro.p2.dual-d2.sd\endcsname{13.2}
\expandafter\def\csname ppt@rev.pol.mdd-tau.sspro.p2.dual-d2.sd\endcsname{$+$13.2}
\expandafter\def\csname val@rev.pol.mdd-tau.sspro.p2.grid2-d3.G\endcsname{12}
\expandafter\def\csname cnt@rev.pol.mdd-tau.sspro.p2.grid2-d3.G\endcsname{12}
\expandafter\def\csname pct@rev.pol.mdd-tau.sspro.p2.grid2-d3.G\endcsname{1200.0}
\expandafter\def\csname ppt@rev.pol.mdd-tau.sspro.p2.grid2-d3.G\endcsname{$+$1200.0}
\expandafter\def\csname val@rev.pol.mdd-tau.sspro.p2.grid2-d3.detectable\endcsname{1}
\expandafter\def\csname cnt@rev.pol.mdd-tau.sspro.p2.grid2-d3.detectable\endcsname{1}
\expandafter\def\csname pct@rev.pol.mdd-tau.sspro.p2.grid2-d3.detectable\endcsname{100.0}
\expandafter\def\csname ppt@rev.pol.mdd-tau.sspro.p2.grid2-d3.detectable\endcsname{$+$100.0}
\expandafter\def\csname val@rev.pol.mdd-tau.sspro.p2.grid2-d3.mdd\endcsname{0.122}
\expandafter\def\csname cnt@rev.pol.mdd-tau.sspro.p2.grid2-d3.mdd\endcsname{0}
\expandafter\def\csname pct@rev.pol.mdd-tau.sspro.p2.grid2-d3.mdd\endcsname{12.2}
\expandafter\def\csname ppt@rev.pol.mdd-tau.sspro.p2.grid2-d3.mdd\endcsname{$+$12.2}
\expandafter\def\csname val@rev.pol.mdd-tau.sspro.p2.grid2-d3.obs\endcsname{0.146}
\expandafter\def\csname cnt@rev.pol.mdd-tau.sspro.p2.grid2-d3.obs\endcsname{0}
\expandafter\def\csname pct@rev.pol.mdd-tau.sspro.p2.grid2-d3.obs\endcsname{14.6}
\expandafter\def\csname ppt@rev.pol.mdd-tau.sspro.p2.grid2-d3.obs\endcsname{$+$14.6}
\expandafter\def\csname val@rev.pol.mdd-tau.sspro.p2.grid2-d3.sd\endcsname{0.158}
\expandafter\def\csname cnt@rev.pol.mdd-tau.sspro.p2.grid2-d3.sd\endcsname{0}
\expandafter\def\csname pct@rev.pol.mdd-tau.sspro.p2.grid2-d3.sd\endcsname{15.8}
\expandafter\def\csname ppt@rev.pol.mdd-tau.sspro.p2.grid2-d3.sd\endcsname{$+$15.8}
\expandafter\def\csname val@rev.pol.mdd-tau.sspro.p2.zoomclick-d3.G\endcsname{12}
\expandafter\def\csname cnt@rev.pol.mdd-tau.sspro.p2.zoomclick-d3.G\endcsname{12}
\expandafter\def\csname pct@rev.pol.mdd-tau.sspro.p2.zoomclick-d3.G\endcsname{1200.0}
\expandafter\def\csname ppt@rev.pol.mdd-tau.sspro.p2.zoomclick-d3.G\endcsname{$+$1200.0}
\expandafter\def\csname val@rev.pol.mdd-tau.sspro.p2.zoomclick-d3.detectable\endcsname{1}
\expandafter\def\csname cnt@rev.pol.mdd-tau.sspro.p2.zoomclick-d3.detectable\endcsname{1}
\expandafter\def\csname pct@rev.pol.mdd-tau.sspro.p2.zoomclick-d3.detectable\endcsname{100.0}
\expandafter\def\csname ppt@rev.pol.mdd-tau.sspro.p2.zoomclick-d3.detectable\endcsname{$+$100.0}
\expandafter\def\csname val@rev.pol.mdd-tau.sspro.p2.zoomclick-d3.mdd\endcsname{0.095}
\expandafter\def\csname cnt@rev.pol.mdd-tau.sspro.p2.zoomclick-d3.mdd\endcsname{0}
\expandafter\def\csname pct@rev.pol.mdd-tau.sspro.p2.zoomclick-d3.mdd\endcsname{9.5}
\expandafter\def\csname ppt@rev.pol.mdd-tau.sspro.p2.zoomclick-d3.mdd\endcsname{$+$9.5}
\expandafter\def\csname val@rev.pol.mdd-tau.sspro.p2.zoomclick-d3.obs\endcsname{0.167}
\expandafter\def\csname cnt@rev.pol.mdd-tau.sspro.p2.zoomclick-d3.obs\endcsname{0}
\expandafter\def\csname pct@rev.pol.mdd-tau.sspro.p2.zoomclick-d3.obs\endcsname{16.7}
\expandafter\def\csname ppt@rev.pol.mdd-tau.sspro.p2.zoomclick-d3.obs\endcsname{$+$16.7}
\expandafter\def\csname val@rev.pol.mdd-tau.sspro.p2.zoomclick-d3.sd\endcsname{0.123}
\expandafter\def\csname cnt@rev.pol.mdd-tau.sspro.p2.zoomclick-d3.sd\endcsname{0}
\expandafter\def\csname pct@rev.pol.mdd-tau.sspro.p2.zoomclick-d3.sd\endcsname{12.3}
\expandafter\def\csname ppt@rev.pol.mdd-tau.sspro.p2.zoomclick-d3.sd\endcsname{$+$12.3}
\expandafter\def\csname val@rev.pol.mdd-tau.uiv.max\endcsname{0.081}
\expandafter\def\csname cnt@rev.pol.mdd-tau.uiv.max\endcsname{0}
\expandafter\def\csname pct@rev.pol.mdd-tau.uiv.max\endcsname{8.1}
\expandafter\def\csname ppt@rev.pol.mdd-tau.uiv.max\endcsname{$+$8.1}
\expandafter\def\csname val@rev.pol.mdd-tau.uiv.min\endcsname{0.036}
\expandafter\def\csname cnt@rev.pol.mdd-tau.uiv.min\endcsname{0}
\expandafter\def\csname pct@rev.pol.mdd-tau.uiv.min\endcsname{3.6}
\expandafter\def\csname ppt@rev.pol.mdd-tau.uiv.min\endcsname{$+$3.6}
\expandafter\def\csname val@rev.pol.mdd-tau.uiv.n\endcsname{8}
\expandafter\def\csname cnt@rev.pol.mdd-tau.uiv.n\endcsname{8}
\expandafter\def\csname pct@rev.pol.mdd-tau.uiv.n\endcsname{800.0}
\expandafter\def\csname ppt@rev.pol.mdd-tau.uiv.n\endcsname{$+$800.0}
\expandafter\def\csname val@rev.pol.mdd-tau.uiv.n-detectable\endcsname{1}
\expandafter\def\csname cnt@rev.pol.mdd-tau.uiv.n-detectable\endcsname{1}
\expandafter\def\csname pct@rev.pol.mdd-tau.uiv.n-detectable\endcsname{100.0}
\expandafter\def\csname ppt@rev.pol.mdd-tau.uiv.n-detectable\endcsname{$+$100.0}
\expandafter\def\csname val@rev.pol.mdd-tau.uiv.p1.center-d1.G\endcsname{24}
\expandafter\def\csname cnt@rev.pol.mdd-tau.uiv.p1.center-d1.G\endcsname{24}
\expandafter\def\csname pct@rev.pol.mdd-tau.uiv.p1.center-d1.G\endcsname{2400.0}
\expandafter\def\csname ppt@rev.pol.mdd-tau.uiv.p1.center-d1.G\endcsname{$+$2400.0}
\expandafter\def\csname val@rev.pol.mdd-tau.uiv.p1.center-d1.detectable\endcsname{0}
\expandafter\def\csname cnt@rev.pol.mdd-tau.uiv.p1.center-d1.detectable\endcsname{0}
\expandafter\def\csname pct@rev.pol.mdd-tau.uiv.p1.center-d1.detectable\endcsname{0.0}
\expandafter\def\csname ppt@rev.pol.mdd-tau.uiv.p1.center-d1.detectable\endcsname{0.0}
\expandafter\def\csname val@rev.pol.mdd-tau.uiv.p1.center-d1.mdd\endcsname{0.080}
\expandafter\def\csname cnt@rev.pol.mdd-tau.uiv.p1.center-d1.mdd\endcsname{0}
\expandafter\def\csname pct@rev.pol.mdd-tau.uiv.p1.center-d1.mdd\endcsname{8.0}
\expandafter\def\csname ppt@rev.pol.mdd-tau.uiv.p1.center-d1.mdd\endcsname{$+$8.0}
\expandafter\def\csname val@rev.pol.mdd-tau.uiv.p1.center-d1.obs\endcsname{-0.056}
\expandafter\def\csname cnt@rev.pol.mdd-tau.uiv.p1.center-d1.obs\endcsname{0}
\expandafter\def\csname pct@rev.pol.mdd-tau.uiv.p1.center-d1.obs\endcsname{$-$5.6}
\expandafter\def\csname ppt@rev.pol.mdd-tau.uiv.p1.center-d1.obs\endcsname{$-$5.6}
\expandafter\def\csname val@rev.pol.mdd-tau.uiv.p1.center-d1.sd\endcsname{0.153}
\expandafter\def\csname cnt@rev.pol.mdd-tau.uiv.p1.center-d1.sd\endcsname{0}
\expandafter\def\csname pct@rev.pol.mdd-tau.uiv.p1.center-d1.sd\endcsname{15.3}
\expandafter\def\csname ppt@rev.pol.mdd-tau.uiv.p1.center-d1.sd\endcsname{$+$15.3}
\expandafter\def\csname val@rev.pol.mdd-tau.uiv.p1.context-d1.G\endcsname{24}
\expandafter\def\csname cnt@rev.pol.mdd-tau.uiv.p1.context-d1.G\endcsname{24}
\expandafter\def\csname pct@rev.pol.mdd-tau.uiv.p1.context-d1.G\endcsname{2400.0}
\expandafter\def\csname ppt@rev.pol.mdd-tau.uiv.p1.context-d1.G\endcsname{$+$2400.0}
\expandafter\def\csname val@rev.pol.mdd-tau.uiv.p1.context-d1.detectable\endcsname{0}
\expandafter\def\csname cnt@rev.pol.mdd-tau.uiv.p1.context-d1.detectable\endcsname{0}
\expandafter\def\csname pct@rev.pol.mdd-tau.uiv.p1.context-d1.detectable\endcsname{0.0}
\expandafter\def\csname ppt@rev.pol.mdd-tau.uiv.p1.context-d1.detectable\endcsname{0.0}
\expandafter\def\csname val@rev.pol.mdd-tau.uiv.p1.context-d1.mdd\endcsname{0.071}
\expandafter\def\csname cnt@rev.pol.mdd-tau.uiv.p1.context-d1.mdd\endcsname{0}
\expandafter\def\csname pct@rev.pol.mdd-tau.uiv.p1.context-d1.mdd\endcsname{7.1}
\expandafter\def\csname ppt@rev.pol.mdd-tau.uiv.p1.context-d1.mdd\endcsname{$+$7.1}
\expandafter\def\csname val@rev.pol.mdd-tau.uiv.p1.context-d1.obs\endcsname{-0.056}
\expandafter\def\csname cnt@rev.pol.mdd-tau.uiv.p1.context-d1.obs\endcsname{0}
\expandafter\def\csname pct@rev.pol.mdd-tau.uiv.p1.context-d1.obs\endcsname{$-$5.6}
\expandafter\def\csname ppt@rev.pol.mdd-tau.uiv.p1.context-d1.obs\endcsname{$-$5.6}
\expandafter\def\csname val@rev.pol.mdd-tau.uiv.p1.context-d1.sd\endcsname{0.136}
\expandafter\def\csname cnt@rev.pol.mdd-tau.uiv.p1.context-d1.sd\endcsname{0}
\expandafter\def\csname pct@rev.pol.mdd-tau.uiv.p1.context-d1.sd\endcsname{13.6}
\expandafter\def\csname ppt@rev.pol.mdd-tau.uiv.p1.context-d1.sd\endcsname{$+$13.6}
\expandafter\def\csname val@rev.pol.mdd-tau.uiv.p1.dual-d2.G\endcsname{24}
\expandafter\def\csname cnt@rev.pol.mdd-tau.uiv.p1.dual-d2.G\endcsname{24}
\expandafter\def\csname pct@rev.pol.mdd-tau.uiv.p1.dual-d2.G\endcsname{2400.0}
\expandafter\def\csname ppt@rev.pol.mdd-tau.uiv.p1.dual-d2.G\endcsname{$+$2400.0}
\expandafter\def\csname val@rev.pol.mdd-tau.uiv.p1.dual-d2.detectable\endcsname{1}
\expandafter\def\csname cnt@rev.pol.mdd-tau.uiv.p1.dual-d2.detectable\endcsname{1}
\expandafter\def\csname pct@rev.pol.mdd-tau.uiv.p1.dual-d2.detectable\endcsname{100.0}
\expandafter\def\csname ppt@rev.pol.mdd-tau.uiv.p1.dual-d2.detectable\endcsname{$+$100.0}
\expandafter\def\csname val@rev.pol.mdd-tau.uiv.p1.dual-d2.mdd\endcsname{0.072}
\expandafter\def\csname cnt@rev.pol.mdd-tau.uiv.p1.dual-d2.mdd\endcsname{0}
\expandafter\def\csname pct@rev.pol.mdd-tau.uiv.p1.dual-d2.mdd\endcsname{7.2}
\expandafter\def\csname ppt@rev.pol.mdd-tau.uiv.p1.dual-d2.mdd\endcsname{$+$7.2}
\expandafter\def\csname val@rev.pol.mdd-tau.uiv.p1.dual-d2.obs\endcsname{-0.104}
\expandafter\def\csname cnt@rev.pol.mdd-tau.uiv.p1.dual-d2.obs\endcsname{0}
\expandafter\def\csname pct@rev.pol.mdd-tau.uiv.p1.dual-d2.obs\endcsname{$-$10.4}
\expandafter\def\csname ppt@rev.pol.mdd-tau.uiv.p1.dual-d2.obs\endcsname{$-$10.4}
\expandafter\def\csname val@rev.pol.mdd-tau.uiv.p1.dual-d2.sd\endcsname{0.137}
\expandafter\def\csname cnt@rev.pol.mdd-tau.uiv.p1.dual-d2.sd\endcsname{0}
\expandafter\def\csname pct@rev.pol.mdd-tau.uiv.p1.dual-d2.sd\endcsname{13.7}
\expandafter\def\csname ppt@rev.pol.mdd-tau.uiv.p1.dual-d2.sd\endcsname{$+$13.7}
\expandafter\def\csname val@rev.pol.mdd-tau.uiv.p1.grid2-d3.G\endcsname{24}
\expandafter\def\csname cnt@rev.pol.mdd-tau.uiv.p1.grid2-d3.G\endcsname{24}
\expandafter\def\csname pct@rev.pol.mdd-tau.uiv.p1.grid2-d3.G\endcsname{2400.0}
\expandafter\def\csname ppt@rev.pol.mdd-tau.uiv.p1.grid2-d3.G\endcsname{$+$2400.0}
\expandafter\def\csname val@rev.pol.mdd-tau.uiv.p1.grid2-d3.detectable\endcsname{0}
\expandafter\def\csname cnt@rev.pol.mdd-tau.uiv.p1.grid2-d3.detectable\endcsname{0}
\expandafter\def\csname pct@rev.pol.mdd-tau.uiv.p1.grid2-d3.detectable\endcsname{0.0}
\expandafter\def\csname ppt@rev.pol.mdd-tau.uiv.p1.grid2-d3.detectable\endcsname{0.0}
\expandafter\def\csname val@rev.pol.mdd-tau.uiv.p1.grid2-d3.mdd\endcsname{0.081}
\expandafter\def\csname cnt@rev.pol.mdd-tau.uiv.p1.grid2-d3.mdd\endcsname{0}
\expandafter\def\csname pct@rev.pol.mdd-tau.uiv.p1.grid2-d3.mdd\endcsname{8.1}
\expandafter\def\csname ppt@rev.pol.mdd-tau.uiv.p1.grid2-d3.mdd\endcsname{$+$8.1}
\expandafter\def\csname val@rev.pol.mdd-tau.uiv.p1.grid2-d3.obs\endcsname{-0.076}
\expandafter\def\csname cnt@rev.pol.mdd-tau.uiv.p1.grid2-d3.obs\endcsname{0}
\expandafter\def\csname pct@rev.pol.mdd-tau.uiv.p1.grid2-d3.obs\endcsname{$-$7.6}
\expandafter\def\csname ppt@rev.pol.mdd-tau.uiv.p1.grid2-d3.obs\endcsname{$-$7.6}
\expandafter\def\csname val@rev.pol.mdd-tau.uiv.p1.grid2-d3.sd\endcsname{0.155}
\expandafter\def\csname cnt@rev.pol.mdd-tau.uiv.p1.grid2-d3.sd\endcsname{0}
\expandafter\def\csname pct@rev.pol.mdd-tau.uiv.p1.grid2-d3.sd\endcsname{15.5}
\expandafter\def\csname ppt@rev.pol.mdd-tau.uiv.p1.grid2-d3.sd\endcsname{$+$15.5}
\expandafter\def\csname val@rev.pol.mdd-tau.uiv.p1.zoomclick-d3.G\endcsname{24}
\expandafter\def\csname cnt@rev.pol.mdd-tau.uiv.p1.zoomclick-d3.G\endcsname{24}
\expandafter\def\csname pct@rev.pol.mdd-tau.uiv.p1.zoomclick-d3.G\endcsname{2400.0}
\expandafter\def\csname ppt@rev.pol.mdd-tau.uiv.p1.zoomclick-d3.G\endcsname{$+$2400.0}
\expandafter\def\csname val@rev.pol.mdd-tau.uiv.p1.zoomclick-d3.detectable\endcsname{0}
\expandafter\def\csname cnt@rev.pol.mdd-tau.uiv.p1.zoomclick-d3.detectable\endcsname{0}
\expandafter\def\csname pct@rev.pol.mdd-tau.uiv.p1.zoomclick-d3.detectable\endcsname{0.0}
\expandafter\def\csname ppt@rev.pol.mdd-tau.uiv.p1.zoomclick-d3.detectable\endcsname{0.0}
\expandafter\def\csname val@rev.pol.mdd-tau.uiv.p1.zoomclick-d3.mdd\endcsname{0.071}
\expandafter\def\csname cnt@rev.pol.mdd-tau.uiv.p1.zoomclick-d3.mdd\endcsname{0}
\expandafter\def\csname pct@rev.pol.mdd-tau.uiv.p1.zoomclick-d3.mdd\endcsname{7.1}
\expandafter\def\csname ppt@rev.pol.mdd-tau.uiv.p1.zoomclick-d3.mdd\endcsname{$+$7.1}
\expandafter\def\csname val@rev.pol.mdd-tau.uiv.p1.zoomclick-d3.obs\endcsname{-0.056}
\expandafter\def\csname cnt@rev.pol.mdd-tau.uiv.p1.zoomclick-d3.obs\endcsname{0}
\expandafter\def\csname pct@rev.pol.mdd-tau.uiv.p1.zoomclick-d3.obs\endcsname{$-$5.6}
\expandafter\def\csname ppt@rev.pol.mdd-tau.uiv.p1.zoomclick-d3.obs\endcsname{$-$5.6}
\expandafter\def\csname val@rev.pol.mdd-tau.uiv.p1.zoomclick-d3.sd\endcsname{0.136}
\expandafter\def\csname cnt@rev.pol.mdd-tau.uiv.p1.zoomclick-d3.sd\endcsname{0}
\expandafter\def\csname pct@rev.pol.mdd-tau.uiv.p1.zoomclick-d3.sd\endcsname{13.6}
\expandafter\def\csname ppt@rev.pol.mdd-tau.uiv.p1.zoomclick-d3.sd\endcsname{$+$13.6}
\expandafter\def\csname val@rev.pol.mdd-tau.uiv.p2.center-d1.G\endcsname{24}
\expandafter\def\csname cnt@rev.pol.mdd-tau.uiv.p2.center-d1.G\endcsname{24}
\expandafter\def\csname pct@rev.pol.mdd-tau.uiv.p2.center-d1.G\endcsname{2400.0}
\expandafter\def\csname ppt@rev.pol.mdd-tau.uiv.p2.center-d1.G\endcsname{$+$2400.0}
\expandafter\def\csname val@rev.pol.mdd-tau.uiv.p2.center-d1.detectable\endcsname{0}
\expandafter\def\csname cnt@rev.pol.mdd-tau.uiv.p2.center-d1.detectable\endcsname{0}
\expandafter\def\csname pct@rev.pol.mdd-tau.uiv.p2.center-d1.detectable\endcsname{0.0}
\expandafter\def\csname ppt@rev.pol.mdd-tau.uiv.p2.center-d1.detectable\endcsname{0.0}
\expandafter\def\csname val@rev.pol.mdd-tau.uiv.p2.center-d1.mdd\endcsname{0.056}
\expandafter\def\csname cnt@rev.pol.mdd-tau.uiv.p2.center-d1.mdd\endcsname{0}
\expandafter\def\csname pct@rev.pol.mdd-tau.uiv.p2.center-d1.mdd\endcsname{5.6}
\expandafter\def\csname ppt@rev.pol.mdd-tau.uiv.p2.center-d1.mdd\endcsname{$+$5.6}
\expandafter\def\csname val@rev.pol.mdd-tau.uiv.p2.center-d1.obs\endcsname{-0.028}
\expandafter\def\csname cnt@rev.pol.mdd-tau.uiv.p2.center-d1.obs\endcsname{0}
\expandafter\def\csname pct@rev.pol.mdd-tau.uiv.p2.center-d1.obs\endcsname{$-$2.8}
\expandafter\def\csname ppt@rev.pol.mdd-tau.uiv.p2.center-d1.obs\endcsname{$-$2.8}
\expandafter\def\csname val@rev.pol.mdd-tau.uiv.p2.center-d1.sd\endcsname{0.106}
\expandafter\def\csname cnt@rev.pol.mdd-tau.uiv.p2.center-d1.sd\endcsname{0}
\expandafter\def\csname pct@rev.pol.mdd-tau.uiv.p2.center-d1.sd\endcsname{10.6}
\expandafter\def\csname ppt@rev.pol.mdd-tau.uiv.p2.center-d1.sd\endcsname{$+$10.6}
\expandafter\def\csname val@rev.pol.mdd-tau.uiv.p2.context-d1.G\endcsname{24}
\expandafter\def\csname cnt@rev.pol.mdd-tau.uiv.p2.context-d1.G\endcsname{24}
\expandafter\def\csname pct@rev.pol.mdd-tau.uiv.p2.context-d1.G\endcsname{2400.0}
\expandafter\def\csname ppt@rev.pol.mdd-tau.uiv.p2.context-d1.G\endcsname{$+$2400.0}
\expandafter\def\csname val@rev.pol.mdd-tau.uiv.p2.context-d1.detectable\endcsname{0}
\expandafter\def\csname cnt@rev.pol.mdd-tau.uiv.p2.context-d1.detectable\endcsname{0}
\expandafter\def\csname pct@rev.pol.mdd-tau.uiv.p2.context-d1.detectable\endcsname{0.0}
\expandafter\def\csname ppt@rev.pol.mdd-tau.uiv.p2.context-d1.detectable\endcsname{0.0}
\expandafter\def\csname val@rev.pol.mdd-tau.uiv.p2.context-d1.mdd\endcsname{0.036}
\expandafter\def\csname cnt@rev.pol.mdd-tau.uiv.p2.context-d1.mdd\endcsname{0}
\expandafter\def\csname pct@rev.pol.mdd-tau.uiv.p2.context-d1.mdd\endcsname{3.6}
\expandafter\def\csname ppt@rev.pol.mdd-tau.uiv.p2.context-d1.mdd\endcsname{$+$3.6}
\expandafter\def\csname val@rev.pol.mdd-tau.uiv.p2.context-d1.obs\endcsname{0.000}
\expandafter\def\csname cnt@rev.pol.mdd-tau.uiv.p2.context-d1.obs\endcsname{0}
\expandafter\def\csname pct@rev.pol.mdd-tau.uiv.p2.context-d1.obs\endcsname{0.0}
\expandafter\def\csname ppt@rev.pol.mdd-tau.uiv.p2.context-d1.obs\endcsname{0.0}
\expandafter\def\csname val@rev.pol.mdd-tau.uiv.p2.context-d1.sd\endcsname{0.070}
\expandafter\def\csname cnt@rev.pol.mdd-tau.uiv.p2.context-d1.sd\endcsname{0}
\expandafter\def\csname pct@rev.pol.mdd-tau.uiv.p2.context-d1.sd\endcsname{7.0}
\expandafter\def\csname ppt@rev.pol.mdd-tau.uiv.p2.context-d1.sd\endcsname{$+$7.0}
\expandafter\def\csname val@rev.pol.mdd-tau.uiv.p2.dual-d2.G\endcsname{24}
\expandafter\def\csname cnt@rev.pol.mdd-tau.uiv.p2.dual-d2.G\endcsname{24}
\expandafter\def\csname pct@rev.pol.mdd-tau.uiv.p2.dual-d2.G\endcsname{2400.0}
\expandafter\def\csname ppt@rev.pol.mdd-tau.uiv.p2.dual-d2.G\endcsname{$+$2400.0}
\expandafter\def\csname val@rev.pol.mdd-tau.uiv.p2.dual-d2.detectable\endcsname{0}
\expandafter\def\csname cnt@rev.pol.mdd-tau.uiv.p2.dual-d2.detectable\endcsname{0}
\expandafter\def\csname pct@rev.pol.mdd-tau.uiv.p2.dual-d2.detectable\endcsname{0.0}
\expandafter\def\csname ppt@rev.pol.mdd-tau.uiv.p2.dual-d2.detectable\endcsname{0.0}
\expandafter\def\csname val@rev.pol.mdd-tau.uiv.p2.dual-d2.mdd\endcsname{0.081}
\expandafter\def\csname cnt@rev.pol.mdd-tau.uiv.p2.dual-d2.mdd\endcsname{0}
\expandafter\def\csname pct@rev.pol.mdd-tau.uiv.p2.dual-d2.mdd\endcsname{8.1}
\expandafter\def\csname ppt@rev.pol.mdd-tau.uiv.p2.dual-d2.mdd\endcsname{$+$8.1}
\expandafter\def\csname val@rev.pol.mdd-tau.uiv.p2.dual-d2.obs\endcsname{-0.062}
\expandafter\def\csname cnt@rev.pol.mdd-tau.uiv.p2.dual-d2.obs\endcsname{0}
\expandafter\def\csname pct@rev.pol.mdd-tau.uiv.p2.dual-d2.obs\endcsname{$-$6.2}
\expandafter\def\csname ppt@rev.pol.mdd-tau.uiv.p2.dual-d2.obs\endcsname{$-$6.2}
\expandafter\def\csname val@rev.pol.mdd-tau.uiv.p2.dual-d2.sd\endcsname{0.154}
\expandafter\def\csname cnt@rev.pol.mdd-tau.uiv.p2.dual-d2.sd\endcsname{0}
\expandafter\def\csname pct@rev.pol.mdd-tau.uiv.p2.dual-d2.sd\endcsname{15.4}
\expandafter\def\csname ppt@rev.pol.mdd-tau.uiv.p2.dual-d2.sd\endcsname{$+$15.4}
\expandafter\def\csname val@rev.pol.mdd-tau.uiv.p2.grid2-d3.G\endcsname{24}
\expandafter\def\csname cnt@rev.pol.mdd-tau.uiv.p2.grid2-d3.G\endcsname{24}
\expandafter\def\csname pct@rev.pol.mdd-tau.uiv.p2.grid2-d3.G\endcsname{2400.0}
\expandafter\def\csname ppt@rev.pol.mdd-tau.uiv.p2.grid2-d3.G\endcsname{$+$2400.0}
\expandafter\def\csname val@rev.pol.mdd-tau.uiv.p2.grid2-d3.detectable\endcsname{0}
\expandafter\def\csname cnt@rev.pol.mdd-tau.uiv.p2.grid2-d3.detectable\endcsname{0}
\expandafter\def\csname pct@rev.pol.mdd-tau.uiv.p2.grid2-d3.detectable\endcsname{0.0}
\expandafter\def\csname ppt@rev.pol.mdd-tau.uiv.p2.grid2-d3.detectable\endcsname{0.0}
\expandafter\def\csname val@rev.pol.mdd-tau.uiv.p2.grid2-d3.mdd\endcsname{0.048}
\expandafter\def\csname cnt@rev.pol.mdd-tau.uiv.p2.grid2-d3.mdd\endcsname{0}
\expandafter\def\csname pct@rev.pol.mdd-tau.uiv.p2.grid2-d3.mdd\endcsname{4.8}
\expandafter\def\csname ppt@rev.pol.mdd-tau.uiv.p2.grid2-d3.mdd\endcsname{$+$4.8}
\expandafter\def\csname val@rev.pol.mdd-tau.uiv.p2.grid2-d3.obs\endcsname{-0.007}
\expandafter\def\csname cnt@rev.pol.mdd-tau.uiv.p2.grid2-d3.obs\endcsname{0}
\expandafter\def\csname pct@rev.pol.mdd-tau.uiv.p2.grid2-d3.obs\endcsname{$-$0.7}
\expandafter\def\csname ppt@rev.pol.mdd-tau.uiv.p2.grid2-d3.obs\endcsname{$-$0.7}
\expandafter\def\csname val@rev.pol.mdd-tau.uiv.p2.grid2-d3.sd\endcsname{0.092}
\expandafter\def\csname cnt@rev.pol.mdd-tau.uiv.p2.grid2-d3.sd\endcsname{0}
\expandafter\def\csname pct@rev.pol.mdd-tau.uiv.p2.grid2-d3.sd\endcsname{9.2}
\expandafter\def\csname ppt@rev.pol.mdd-tau.uiv.p2.grid2-d3.sd\endcsname{$+$9.2}
\expandafter\def\csname val@rev.pol.mdd-tau.uiv.p2.zoomclick-d3.G\endcsname{24}
\expandafter\def\csname cnt@rev.pol.mdd-tau.uiv.p2.zoomclick-d3.G\endcsname{24}
\expandafter\def\csname pct@rev.pol.mdd-tau.uiv.p2.zoomclick-d3.G\endcsname{2400.0}
\expandafter\def\csname ppt@rev.pol.mdd-tau.uiv.p2.zoomclick-d3.G\endcsname{$+$2400.0}
\expandafter\def\csname val@rev.pol.mdd-tau.uiv.p2.zoomclick-d3.detectable\endcsname{0}
\expandafter\def\csname cnt@rev.pol.mdd-tau.uiv.p2.zoomclick-d3.detectable\endcsname{0}
\expandafter\def\csname pct@rev.pol.mdd-tau.uiv.p2.zoomclick-d3.detectable\endcsname{0.0}
\expandafter\def\csname ppt@rev.pol.mdd-tau.uiv.p2.zoomclick-d3.detectable\endcsname{0.0}
\expandafter\def\csname val@rev.pol.mdd-tau.uiv.p2.zoomclick-d3.mdd\endcsname{0.055}
\expandafter\def\csname cnt@rev.pol.mdd-tau.uiv.p2.zoomclick-d3.mdd\endcsname{0}
\expandafter\def\csname pct@rev.pol.mdd-tau.uiv.p2.zoomclick-d3.mdd\endcsname{5.5}
\expandafter\def\csname ppt@rev.pol.mdd-tau.uiv.p2.zoomclick-d3.mdd\endcsname{$+$5.5}
\expandafter\def\csname val@rev.pol.mdd-tau.uiv.p2.zoomclick-d3.obs\endcsname{-0.007}
\expandafter\def\csname cnt@rev.pol.mdd-tau.uiv.p2.zoomclick-d3.obs\endcsname{0}
\expandafter\def\csname pct@rev.pol.mdd-tau.uiv.p2.zoomclick-d3.obs\endcsname{$-$0.7}
\expandafter\def\csname ppt@rev.pol.mdd-tau.uiv.p2.zoomclick-d3.obs\endcsname{$-$0.7}
\expandafter\def\csname val@rev.pol.mdd-tau.uiv.p2.zoomclick-d3.sd\endcsname{0.104}
\expandafter\def\csname cnt@rev.pol.mdd-tau.uiv.p2.zoomclick-d3.sd\endcsname{0}
\expandafter\def\csname pct@rev.pol.mdd-tau.uiv.p2.zoomclick-d3.sd\endcsname{10.4}
\expandafter\def\csname ppt@rev.pol.mdd-tau.uiv.p2.zoomclick-d3.sd\endcsname{$+$10.4}
\expandafter\def\csname val@rev.pol.mdd.f20.sspro.p1.rcr-vs-full-recert-ni.G\endcsname{12}
\expandafter\def\csname cnt@rev.pol.mdd.f20.sspro.p1.rcr-vs-full-recert-ni.G\endcsname{12}
\expandafter\def\csname pct@rev.pol.mdd.f20.sspro.p1.rcr-vs-full-recert-ni.G\endcsname{1200.0}
\expandafter\def\csname ppt@rev.pol.mdd.f20.sspro.p1.rcr-vs-full-recert-ni.G\endcsname{$+$1200.0}
\expandafter\def\csname val@rev.pol.mdd.f20.sspro.p1.rcr-vs-full-recert-ni.mdd\endcsname{0.003}
\expandafter\def\csname cnt@rev.pol.mdd.f20.sspro.p1.rcr-vs-full-recert-ni.mdd\endcsname{0}
\expandafter\def\csname pct@rev.pol.mdd.f20.sspro.p1.rcr-vs-full-recert-ni.mdd\endcsname{0.3}
\expandafter\def\csname ppt@rev.pol.mdd.f20.sspro.p1.rcr-vs-full-recert-ni.mdd\endcsname{$+$0.3}
\expandafter\def\csname val@rev.pol.mdd.f20.sspro.p1.rcr-vs-full-recert-ni.mdd-holm\endcsname{0.004}
\expandafter\def\csname cnt@rev.pol.mdd.f20.sspro.p1.rcr-vs-full-recert-ni.mdd-holm\endcsname{0}
\expandafter\def\csname pct@rev.pol.mdd.f20.sspro.p1.rcr-vs-full-recert-ni.mdd-holm\endcsname{0.4}
\expandafter\def\csname ppt@rev.pol.mdd.f20.sspro.p1.rcr-vs-full-recert-ni.mdd-holm\endcsname{$+$0.4}
\expandafter\def\csname val@rev.pol.mdd.f20.sspro.p1.rcr-vs-full-recert-ni.obs\endcsname{-0.002}
\expandafter\def\csname cnt@rev.pol.mdd.f20.sspro.p1.rcr-vs-full-recert-ni.obs\endcsname{0}
\expandafter\def\csname pct@rev.pol.mdd.f20.sspro.p1.rcr-vs-full-recert-ni.obs\endcsname{$-$0.2}
\expandafter\def\csname ppt@rev.pol.mdd.f20.sspro.p1.rcr-vs-full-recert-ni.obs\endcsname{$-$0.2}
\expandafter\def\csname val@rev.pol.mdd.f20.sspro.p1.rcr-vs-full-recert-ni.sd\endcsname{0.003}
\expandafter\def\csname cnt@rev.pol.mdd.f20.sspro.p1.rcr-vs-full-recert-ni.sd\endcsname{0}
\expandafter\def\csname pct@rev.pol.mdd.f20.sspro.p1.rcr-vs-full-recert-ni.sd\endcsname{0.3}
\expandafter\def\csname ppt@rev.pol.mdd.f20.sspro.p1.rcr-vs-full-recert-ni.sd\endcsname{$+$0.3}
\expandafter\def\csname val@rev.pol.mdd.f20.sspro.p1.rcr-vs-generic-ppi-source.G\endcsname{12}
\expandafter\def\csname cnt@rev.pol.mdd.f20.sspro.p1.rcr-vs-generic-ppi-source.G\endcsname{12}
\expandafter\def\csname pct@rev.pol.mdd.f20.sspro.p1.rcr-vs-generic-ppi-source.G\endcsname{1200.0}
\expandafter\def\csname ppt@rev.pol.mdd.f20.sspro.p1.rcr-vs-generic-ppi-source.G\endcsname{$+$1200.0}
\expandafter\def\csname val@rev.pol.mdd.f20.sspro.p1.rcr-vs-generic-ppi-source.mdd\endcsname{0.008}
\expandafter\def\csname cnt@rev.pol.mdd.f20.sspro.p1.rcr-vs-generic-ppi-source.mdd\endcsname{0}
\expandafter\def\csname pct@rev.pol.mdd.f20.sspro.p1.rcr-vs-generic-ppi-source.mdd\endcsname{0.8}
\expandafter\def\csname ppt@rev.pol.mdd.f20.sspro.p1.rcr-vs-generic-ppi-source.mdd\endcsname{$+$0.8}
\expandafter\def\csname val@rev.pol.mdd.f20.sspro.p1.rcr-vs-generic-ppi-source.mdd-holm\endcsname{0.012}
\expandafter\def\csname cnt@rev.pol.mdd.f20.sspro.p1.rcr-vs-generic-ppi-source.mdd-holm\endcsname{0}
\expandafter\def\csname pct@rev.pol.mdd.f20.sspro.p1.rcr-vs-generic-ppi-source.mdd-holm\endcsname{1.2}
\expandafter\def\csname ppt@rev.pol.mdd.f20.sspro.p1.rcr-vs-generic-ppi-source.mdd-holm\endcsname{$+$1.2}
\expandafter\def\csname val@rev.pol.mdd.f20.sspro.p1.rcr-vs-generic-ppi-source.obs\endcsname{0.006}
\expandafter\def\csname cnt@rev.pol.mdd.f20.sspro.p1.rcr-vs-generic-ppi-source.obs\endcsname{0}
\expandafter\def\csname pct@rev.pol.mdd.f20.sspro.p1.rcr-vs-generic-ppi-source.obs\endcsname{0.6}
\expandafter\def\csname ppt@rev.pol.mdd.f20.sspro.p1.rcr-vs-generic-ppi-source.obs\endcsname{$+$0.6}
\expandafter\def\csname val@rev.pol.mdd.f20.sspro.p1.rcr-vs-generic-ppi-source.sd\endcsname{0.010}
\expandafter\def\csname cnt@rev.pol.mdd.f20.sspro.p1.rcr-vs-generic-ppi-source.sd\endcsname{0}
\expandafter\def\csname pct@rev.pol.mdd.f20.sspro.p1.rcr-vs-generic-ppi-source.sd\endcsname{1.0}
\expandafter\def\csname ppt@rev.pol.mdd.f20.sspro.p1.rcr-vs-generic-ppi-source.sd\endcsname{$+$1.0}
\expandafter\def\csname val@rev.pol.mdd.f20.sspro.p1.rcr-vs-replay-ppi-uniform.G\endcsname{12}
\expandafter\def\csname cnt@rev.pol.mdd.f20.sspro.p1.rcr-vs-replay-ppi-uniform.G\endcsname{12}
\expandafter\def\csname pct@rev.pol.mdd.f20.sspro.p1.rcr-vs-replay-ppi-uniform.G\endcsname{1200.0}
\expandafter\def\csname ppt@rev.pol.mdd.f20.sspro.p1.rcr-vs-replay-ppi-uniform.G\endcsname{$+$1200.0}
\expandafter\def\csname val@rev.pol.mdd.f20.sspro.p1.rcr-vs-replay-ppi-uniform.mdd\endcsname{0.012}
\expandafter\def\csname cnt@rev.pol.mdd.f20.sspro.p1.rcr-vs-replay-ppi-uniform.mdd\endcsname{0}
\expandafter\def\csname pct@rev.pol.mdd.f20.sspro.p1.rcr-vs-replay-ppi-uniform.mdd\endcsname{1.2}
\expandafter\def\csname ppt@rev.pol.mdd.f20.sspro.p1.rcr-vs-replay-ppi-uniform.mdd\endcsname{$+$1.2}
\expandafter\def\csname val@rev.pol.mdd.f20.sspro.p1.rcr-vs-replay-ppi-uniform.mdd-holm\endcsname{0.018}
\expandafter\def\csname cnt@rev.pol.mdd.f20.sspro.p1.rcr-vs-replay-ppi-uniform.mdd-holm\endcsname{0}
\expandafter\def\csname pct@rev.pol.mdd.f20.sspro.p1.rcr-vs-replay-ppi-uniform.mdd-holm\endcsname{1.8}
\expandafter\def\csname ppt@rev.pol.mdd.f20.sspro.p1.rcr-vs-replay-ppi-uniform.mdd-holm\endcsname{$+$1.8}
\expandafter\def\csname val@rev.pol.mdd.f20.sspro.p1.rcr-vs-replay-ppi-uniform.obs\endcsname{0.009}
\expandafter\def\csname cnt@rev.pol.mdd.f20.sspro.p1.rcr-vs-replay-ppi-uniform.obs\endcsname{0}
\expandafter\def\csname pct@rev.pol.mdd.f20.sspro.p1.rcr-vs-replay-ppi-uniform.obs\endcsname{0.9}
\expandafter\def\csname ppt@rev.pol.mdd.f20.sspro.p1.rcr-vs-replay-ppi-uniform.obs\endcsname{$+$0.9}
\expandafter\def\csname val@rev.pol.mdd.f20.sspro.p1.rcr-vs-replay-ppi-uniform.sd\endcsname{0.016}
\expandafter\def\csname cnt@rev.pol.mdd.f20.sspro.p1.rcr-vs-replay-ppi-uniform.sd\endcsname{0}
\expandafter\def\csname pct@rev.pol.mdd.f20.sspro.p1.rcr-vs-replay-ppi-uniform.sd\endcsname{1.6}
\expandafter\def\csname ppt@rev.pol.mdd.f20.sspro.p1.rcr-vs-replay-ppi-uniform.sd\endcsname{$+$1.6}
\expandafter\def\csname val@rev.pol.mdd.f20.sspro.p2.rcr-vs-full-recert-ni.G\endcsname{12}
\expandafter\def\csname cnt@rev.pol.mdd.f20.sspro.p2.rcr-vs-full-recert-ni.G\endcsname{12}
\expandafter\def\csname pct@rev.pol.mdd.f20.sspro.p2.rcr-vs-full-recert-ni.G\endcsname{1200.0}
\expandafter\def\csname ppt@rev.pol.mdd.f20.sspro.p2.rcr-vs-full-recert-ni.G\endcsname{$+$1200.0}
\expandafter\def\csname val@rev.pol.mdd.f20.sspro.p2.rcr-vs-full-recert-ni.mdd\endcsname{0.008}
\expandafter\def\csname cnt@rev.pol.mdd.f20.sspro.p2.rcr-vs-full-recert-ni.mdd\endcsname{0}
\expandafter\def\csname pct@rev.pol.mdd.f20.sspro.p2.rcr-vs-full-recert-ni.mdd\endcsname{0.8}
\expandafter\def\csname ppt@rev.pol.mdd.f20.sspro.p2.rcr-vs-full-recert-ni.mdd\endcsname{$+$0.8}
\expandafter\def\csname val@rev.pol.mdd.f20.sspro.p2.rcr-vs-full-recert-ni.mdd-holm\endcsname{0.011}
\expandafter\def\csname cnt@rev.pol.mdd.f20.sspro.p2.rcr-vs-full-recert-ni.mdd-holm\endcsname{0}
\expandafter\def\csname pct@rev.pol.mdd.f20.sspro.p2.rcr-vs-full-recert-ni.mdd-holm\endcsname{1.1}
\expandafter\def\csname ppt@rev.pol.mdd.f20.sspro.p2.rcr-vs-full-recert-ni.mdd-holm\endcsname{$+$1.1}
\expandafter\def\csname val@rev.pol.mdd.f20.sspro.p2.rcr-vs-full-recert-ni.obs\endcsname{-0.011}
\expandafter\def\csname cnt@rev.pol.mdd.f20.sspro.p2.rcr-vs-full-recert-ni.obs\endcsname{0}
\expandafter\def\csname pct@rev.pol.mdd.f20.sspro.p2.rcr-vs-full-recert-ni.obs\endcsname{$-$1.1}
\expandafter\def\csname ppt@rev.pol.mdd.f20.sspro.p2.rcr-vs-full-recert-ni.obs\endcsname{$-$1.1}
\expandafter\def\csname val@rev.pol.mdd.f20.sspro.p2.rcr-vs-full-recert-ni.sd\endcsname{0.010}
\expandafter\def\csname cnt@rev.pol.mdd.f20.sspro.p2.rcr-vs-full-recert-ni.sd\endcsname{0}
\expandafter\def\csname pct@rev.pol.mdd.f20.sspro.p2.rcr-vs-full-recert-ni.sd\endcsname{1.0}
\expandafter\def\csname ppt@rev.pol.mdd.f20.sspro.p2.rcr-vs-full-recert-ni.sd\endcsname{$+$1.0}
\expandafter\def\csname val@rev.pol.mdd.f20.sspro.p2.rcr-vs-generic-ppi-source.G\endcsname{12}
\expandafter\def\csname cnt@rev.pol.mdd.f20.sspro.p2.rcr-vs-generic-ppi-source.G\endcsname{12}
\expandafter\def\csname pct@rev.pol.mdd.f20.sspro.p2.rcr-vs-generic-ppi-source.G\endcsname{1200.0}
\expandafter\def\csname ppt@rev.pol.mdd.f20.sspro.p2.rcr-vs-generic-ppi-source.G\endcsname{$+$1200.0}
\expandafter\def\csname val@rev.pol.mdd.f20.sspro.p2.rcr-vs-generic-ppi-source.mdd\endcsname{0.006}
\expandafter\def\csname cnt@rev.pol.mdd.f20.sspro.p2.rcr-vs-generic-ppi-source.mdd\endcsname{0}
\expandafter\def\csname pct@rev.pol.mdd.f20.sspro.p2.rcr-vs-generic-ppi-source.mdd\endcsname{0.6}
\expandafter\def\csname ppt@rev.pol.mdd.f20.sspro.p2.rcr-vs-generic-ppi-source.mdd\endcsname{$+$0.6}
\expandafter\def\csname val@rev.pol.mdd.f20.sspro.p2.rcr-vs-generic-ppi-source.mdd-holm\endcsname{0.009}
\expandafter\def\csname cnt@rev.pol.mdd.f20.sspro.p2.rcr-vs-generic-ppi-source.mdd-holm\endcsname{0}
\expandafter\def\csname pct@rev.pol.mdd.f20.sspro.p2.rcr-vs-generic-ppi-source.mdd-holm\endcsname{0.9}
\expandafter\def\csname ppt@rev.pol.mdd.f20.sspro.p2.rcr-vs-generic-ppi-source.mdd-holm\endcsname{$+$0.9}
\expandafter\def\csname val@rev.pol.mdd.f20.sspro.p2.rcr-vs-generic-ppi-source.obs\endcsname{0.010}
\expandafter\def\csname cnt@rev.pol.mdd.f20.sspro.p2.rcr-vs-generic-ppi-source.obs\endcsname{0}
\expandafter\def\csname pct@rev.pol.mdd.f20.sspro.p2.rcr-vs-generic-ppi-source.obs\endcsname{1.0}
\expandafter\def\csname ppt@rev.pol.mdd.f20.sspro.p2.rcr-vs-generic-ppi-source.obs\endcsname{$+$1.0}
\expandafter\def\csname val@rev.pol.mdd.f20.sspro.p2.rcr-vs-generic-ppi-source.sd\endcsname{0.008}
\expandafter\def\csname cnt@rev.pol.mdd.f20.sspro.p2.rcr-vs-generic-ppi-source.sd\endcsname{0}
\expandafter\def\csname pct@rev.pol.mdd.f20.sspro.p2.rcr-vs-generic-ppi-source.sd\endcsname{0.8}
\expandafter\def\csname ppt@rev.pol.mdd.f20.sspro.p2.rcr-vs-generic-ppi-source.sd\endcsname{$+$0.8}
\expandafter\def\csname val@rev.pol.mdd.f20.sspro.p2.rcr-vs-replay-ppi-uniform.G\endcsname{12}
\expandafter\def\csname cnt@rev.pol.mdd.f20.sspro.p2.rcr-vs-replay-ppi-uniform.G\endcsname{12}
\expandafter\def\csname pct@rev.pol.mdd.f20.sspro.p2.rcr-vs-replay-ppi-uniform.G\endcsname{1200.0}
\expandafter\def\csname ppt@rev.pol.mdd.f20.sspro.p2.rcr-vs-replay-ppi-uniform.G\endcsname{$+$1200.0}
\expandafter\def\csname val@rev.pol.mdd.f20.sspro.p2.rcr-vs-replay-ppi-uniform.mdd\endcsname{0.006}
\expandafter\def\csname cnt@rev.pol.mdd.f20.sspro.p2.rcr-vs-replay-ppi-uniform.mdd\endcsname{0}
\expandafter\def\csname pct@rev.pol.mdd.f20.sspro.p2.rcr-vs-replay-ppi-uniform.mdd\endcsname{0.6}
\expandafter\def\csname ppt@rev.pol.mdd.f20.sspro.p2.rcr-vs-replay-ppi-uniform.mdd\endcsname{$+$0.6}
\expandafter\def\csname val@rev.pol.mdd.f20.sspro.p2.rcr-vs-replay-ppi-uniform.mdd-holm\endcsname{0.009}
\expandafter\def\csname cnt@rev.pol.mdd.f20.sspro.p2.rcr-vs-replay-ppi-uniform.mdd-holm\endcsname{0}
\expandafter\def\csname pct@rev.pol.mdd.f20.sspro.p2.rcr-vs-replay-ppi-uniform.mdd-holm\endcsname{0.9}
\expandafter\def\csname ppt@rev.pol.mdd.f20.sspro.p2.rcr-vs-replay-ppi-uniform.mdd-holm\endcsname{$+$0.9}
\expandafter\def\csname val@rev.pol.mdd.f20.sspro.p2.rcr-vs-replay-ppi-uniform.obs\endcsname{0.010}
\expandafter\def\csname cnt@rev.pol.mdd.f20.sspro.p2.rcr-vs-replay-ppi-uniform.obs\endcsname{0}
\expandafter\def\csname pct@rev.pol.mdd.f20.sspro.p2.rcr-vs-replay-ppi-uniform.obs\endcsname{1.0}
\expandafter\def\csname ppt@rev.pol.mdd.f20.sspro.p2.rcr-vs-replay-ppi-uniform.obs\endcsname{$+$1.0}
\expandafter\def\csname val@rev.pol.mdd.f20.sspro.p2.rcr-vs-replay-ppi-uniform.sd\endcsname{0.008}
\expandafter\def\csname cnt@rev.pol.mdd.f20.sspro.p2.rcr-vs-replay-ppi-uniform.sd\endcsname{0}
\expandafter\def\csname pct@rev.pol.mdd.f20.sspro.p2.rcr-vs-replay-ppi-uniform.sd\endcsname{0.8}
\expandafter\def\csname ppt@rev.pol.mdd.f20.sspro.p2.rcr-vs-replay-ppi-uniform.sd\endcsname{$+$0.8}
\expandafter\def\csname val@rev.pol.mdd.f20.sup.holm-max\endcsname{0.018}
\expandafter\def\csname cnt@rev.pol.mdd.f20.sup.holm-max\endcsname{0}
\expandafter\def\csname pct@rev.pol.mdd.f20.sup.holm-max\endcsname{1.8}
\expandafter\def\csname ppt@rev.pol.mdd.f20.sup.holm-max\endcsname{$+$1.8}
\expandafter\def\csname val@rev.pol.mdd.f20.sup.holm-min\endcsname{0.003}
\expandafter\def\csname cnt@rev.pol.mdd.f20.sup.holm-min\endcsname{0}
\expandafter\def\csname pct@rev.pol.mdd.f20.sup.holm-min\endcsname{0.3}
\expandafter\def\csname ppt@rev.pol.mdd.f20.sup.holm-min\endcsname{$+$0.3}
\expandafter\def\csname val@rev.pol.mdd.f20.sup.max\endcsname{0.012}
\expandafter\def\csname cnt@rev.pol.mdd.f20.sup.max\endcsname{0}
\expandafter\def\csname pct@rev.pol.mdd.f20.sup.max\endcsname{1.2}
\expandafter\def\csname ppt@rev.pol.mdd.f20.sup.max\endcsname{$+$1.2}
\expandafter\def\csname val@rev.pol.mdd.f20.sup.min\endcsname{0.002}
\expandafter\def\csname cnt@rev.pol.mdd.f20.sup.min\endcsname{0}
\expandafter\def\csname pct@rev.pol.mdd.f20.sup.min\endcsname{0.2}
\expandafter\def\csname ppt@rev.pol.mdd.f20.sup.min\endcsname{$+$0.2}
\expandafter\def\csname val@rev.pol.mdd.f20.sup.n-obs-above-mdd\endcsname{3}
\expandafter\def\csname cnt@rev.pol.mdd.f20.sup.n-obs-above-mdd\endcsname{3}
\expandafter\def\csname pct@rev.pol.mdd.f20.sup.n-obs-above-mdd\endcsname{300.0}
\expandafter\def\csname ppt@rev.pol.mdd.f20.sup.n-obs-above-mdd\endcsname{$+$300.0}
\expandafter\def\csname val@rev.pol.mdd.f20.sup.n-obs-above-mdd-holm\endcsname{3}
\expandafter\def\csname cnt@rev.pol.mdd.f20.sup.n-obs-above-mdd-holm\endcsname{3}
\expandafter\def\csname pct@rev.pol.mdd.f20.sup.n-obs-above-mdd-holm\endcsname{300.0}
\expandafter\def\csname ppt@rev.pol.mdd.f20.sup.n-obs-above-mdd-holm\endcsname{$+$300.0}
\expandafter\def\csname val@rev.pol.mdd.f20.uiv.p1.rcr-vs-full-recert-ni.G\endcsname{24}
\expandafter\def\csname cnt@rev.pol.mdd.f20.uiv.p1.rcr-vs-full-recert-ni.G\endcsname{24}
\expandafter\def\csname pct@rev.pol.mdd.f20.uiv.p1.rcr-vs-full-recert-ni.G\endcsname{2400.0}
\expandafter\def\csname ppt@rev.pol.mdd.f20.uiv.p1.rcr-vs-full-recert-ni.G\endcsname{$+$2400.0}
\expandafter\def\csname val@rev.pol.mdd.f20.uiv.p1.rcr-vs-full-recert-ni.mdd\endcsname{0.011}
\expandafter\def\csname cnt@rev.pol.mdd.f20.uiv.p1.rcr-vs-full-recert-ni.mdd\endcsname{0}
\expandafter\def\csname pct@rev.pol.mdd.f20.uiv.p1.rcr-vs-full-recert-ni.mdd\endcsname{1.1}
\expandafter\def\csname ppt@rev.pol.mdd.f20.uiv.p1.rcr-vs-full-recert-ni.mdd\endcsname{$+$1.1}
\expandafter\def\csname val@rev.pol.mdd.f20.uiv.p1.rcr-vs-full-recert-ni.mdd-holm\endcsname{0.015}
\expandafter\def\csname cnt@rev.pol.mdd.f20.uiv.p1.rcr-vs-full-recert-ni.mdd-holm\endcsname{0}
\expandafter\def\csname pct@rev.pol.mdd.f20.uiv.p1.rcr-vs-full-recert-ni.mdd-holm\endcsname{1.5}
\expandafter\def\csname ppt@rev.pol.mdd.f20.uiv.p1.rcr-vs-full-recert-ni.mdd-holm\endcsname{$+$1.5}
\expandafter\def\csname val@rev.pol.mdd.f20.uiv.p1.rcr-vs-full-recert-ni.obs\endcsname{0.010}
\expandafter\def\csname cnt@rev.pol.mdd.f20.uiv.p1.rcr-vs-full-recert-ni.obs\endcsname{0}
\expandafter\def\csname pct@rev.pol.mdd.f20.uiv.p1.rcr-vs-full-recert-ni.obs\endcsname{1.0}
\expandafter\def\csname ppt@rev.pol.mdd.f20.uiv.p1.rcr-vs-full-recert-ni.obs\endcsname{$+$1.0}
\expandafter\def\csname val@rev.pol.mdd.f20.uiv.p1.rcr-vs-full-recert-ni.sd\endcsname{0.020}
\expandafter\def\csname cnt@rev.pol.mdd.f20.uiv.p1.rcr-vs-full-recert-ni.sd\endcsname{0}
\expandafter\def\csname pct@rev.pol.mdd.f20.uiv.p1.rcr-vs-full-recert-ni.sd\endcsname{2.0}
\expandafter\def\csname ppt@rev.pol.mdd.f20.uiv.p1.rcr-vs-full-recert-ni.sd\endcsname{$+$2.0}
\expandafter\def\csname val@rev.pol.mdd.f20.uiv.p1.rcr-vs-generic-ppi-source.G\endcsname{24}
\expandafter\def\csname cnt@rev.pol.mdd.f20.uiv.p1.rcr-vs-generic-ppi-source.G\endcsname{24}
\expandafter\def\csname pct@rev.pol.mdd.f20.uiv.p1.rcr-vs-generic-ppi-source.G\endcsname{2400.0}
\expandafter\def\csname ppt@rev.pol.mdd.f20.uiv.p1.rcr-vs-generic-ppi-source.G\endcsname{$+$2400.0}
\expandafter\def\csname val@rev.pol.mdd.f20.uiv.p1.rcr-vs-generic-ppi-source.mdd\endcsname{0.009}
\expandafter\def\csname cnt@rev.pol.mdd.f20.uiv.p1.rcr-vs-generic-ppi-source.mdd\endcsname{0}
\expandafter\def\csname pct@rev.pol.mdd.f20.uiv.p1.rcr-vs-generic-ppi-source.mdd\endcsname{0.9}
\expandafter\def\csname ppt@rev.pol.mdd.f20.uiv.p1.rcr-vs-generic-ppi-source.mdd\endcsname{$+$0.9}
\expandafter\def\csname val@rev.pol.mdd.f20.uiv.p1.rcr-vs-generic-ppi-source.mdd-holm\endcsname{0.013}
\expandafter\def\csname cnt@rev.pol.mdd.f20.uiv.p1.rcr-vs-generic-ppi-source.mdd-holm\endcsname{0}
\expandafter\def\csname pct@rev.pol.mdd.f20.uiv.p1.rcr-vs-generic-ppi-source.mdd-holm\endcsname{1.3}
\expandafter\def\csname ppt@rev.pol.mdd.f20.uiv.p1.rcr-vs-generic-ppi-source.mdd-holm\endcsname{$+$1.3}
\expandafter\def\csname val@rev.pol.mdd.f20.uiv.p1.rcr-vs-generic-ppi-source.obs\endcsname{0.013}
\expandafter\def\csname cnt@rev.pol.mdd.f20.uiv.p1.rcr-vs-generic-ppi-source.obs\endcsname{0}
\expandafter\def\csname pct@rev.pol.mdd.f20.uiv.p1.rcr-vs-generic-ppi-source.obs\endcsname{1.3}
\expandafter\def\csname ppt@rev.pol.mdd.f20.uiv.p1.rcr-vs-generic-ppi-source.obs\endcsname{$+$1.3}
\expandafter\def\csname val@rev.pol.mdd.f20.uiv.p1.rcr-vs-generic-ppi-source.sd\endcsname{0.017}
\expandafter\def\csname cnt@rev.pol.mdd.f20.uiv.p1.rcr-vs-generic-ppi-source.sd\endcsname{0}
\expandafter\def\csname pct@rev.pol.mdd.f20.uiv.p1.rcr-vs-generic-ppi-source.sd\endcsname{1.7}
\expandafter\def\csname ppt@rev.pol.mdd.f20.uiv.p1.rcr-vs-generic-ppi-source.sd\endcsname{$+$1.7}
\expandafter\def\csname val@rev.pol.mdd.f20.uiv.p1.rcr-vs-replay-ppi-uniform.G\endcsname{24}
\expandafter\def\csname cnt@rev.pol.mdd.f20.uiv.p1.rcr-vs-replay-ppi-uniform.G\endcsname{24}
\expandafter\def\csname pct@rev.pol.mdd.f20.uiv.p1.rcr-vs-replay-ppi-uniform.G\endcsname{2400.0}
\expandafter\def\csname ppt@rev.pol.mdd.f20.uiv.p1.rcr-vs-replay-ppi-uniform.G\endcsname{$+$2400.0}
\expandafter\def\csname val@rev.pol.mdd.f20.uiv.p1.rcr-vs-replay-ppi-uniform.mdd\endcsname{0.004}
\expandafter\def\csname cnt@rev.pol.mdd.f20.uiv.p1.rcr-vs-replay-ppi-uniform.mdd\endcsname{0}
\expandafter\def\csname pct@rev.pol.mdd.f20.uiv.p1.rcr-vs-replay-ppi-uniform.mdd\endcsname{0.4}
\expandafter\def\csname ppt@rev.pol.mdd.f20.uiv.p1.rcr-vs-replay-ppi-uniform.mdd\endcsname{$+$0.4}
\expandafter\def\csname val@rev.pol.mdd.f20.uiv.p1.rcr-vs-replay-ppi-uniform.mdd-holm\endcsname{0.005}
\expandafter\def\csname cnt@rev.pol.mdd.f20.uiv.p1.rcr-vs-replay-ppi-uniform.mdd-holm\endcsname{0}
\expandafter\def\csname pct@rev.pol.mdd.f20.uiv.p1.rcr-vs-replay-ppi-uniform.mdd-holm\endcsname{0.5}
\expandafter\def\csname ppt@rev.pol.mdd.f20.uiv.p1.rcr-vs-replay-ppi-uniform.mdd-holm\endcsname{$+$0.5}
\expandafter\def\csname val@rev.pol.mdd.f20.uiv.p1.rcr-vs-replay-ppi-uniform.obs\endcsname{0.001}
\expandafter\def\csname cnt@rev.pol.mdd.f20.uiv.p1.rcr-vs-replay-ppi-uniform.obs\endcsname{0}
\expandafter\def\csname pct@rev.pol.mdd.f20.uiv.p1.rcr-vs-replay-ppi-uniform.obs\endcsname{0.1}
\expandafter\def\csname ppt@rev.pol.mdd.f20.uiv.p1.rcr-vs-replay-ppi-uniform.obs\endcsname{$+$0.1}
\expandafter\def\csname val@rev.pol.mdd.f20.uiv.p1.rcr-vs-replay-ppi-uniform.sd\endcsname{0.007}
\expandafter\def\csname cnt@rev.pol.mdd.f20.uiv.p1.rcr-vs-replay-ppi-uniform.sd\endcsname{0}
\expandafter\def\csname pct@rev.pol.mdd.f20.uiv.p1.rcr-vs-replay-ppi-uniform.sd\endcsname{0.7}
\expandafter\def\csname ppt@rev.pol.mdd.f20.uiv.p1.rcr-vs-replay-ppi-uniform.sd\endcsname{$+$0.7}
\expandafter\def\csname val@rev.pol.mdd.f20.uiv.p2.rcr-vs-full-recert-ni.G\endcsname{24}
\expandafter\def\csname cnt@rev.pol.mdd.f20.uiv.p2.rcr-vs-full-recert-ni.G\endcsname{24}
\expandafter\def\csname pct@rev.pol.mdd.f20.uiv.p2.rcr-vs-full-recert-ni.G\endcsname{2400.0}
\expandafter\def\csname ppt@rev.pol.mdd.f20.uiv.p2.rcr-vs-full-recert-ni.G\endcsname{$+$2400.0}
\expandafter\def\csname val@rev.pol.mdd.f20.uiv.p2.rcr-vs-full-recert-ni.mdd\endcsname{0.010}
\expandafter\def\csname cnt@rev.pol.mdd.f20.uiv.p2.rcr-vs-full-recert-ni.mdd\endcsname{0}
\expandafter\def\csname pct@rev.pol.mdd.f20.uiv.p2.rcr-vs-full-recert-ni.mdd\endcsname{1.0}
\expandafter\def\csname ppt@rev.pol.mdd.f20.uiv.p2.rcr-vs-full-recert-ni.mdd\endcsname{$+$1.0}
\expandafter\def\csname val@rev.pol.mdd.f20.uiv.p2.rcr-vs-full-recert-ni.mdd-holm\endcsname{0.014}
\expandafter\def\csname cnt@rev.pol.mdd.f20.uiv.p2.rcr-vs-full-recert-ni.mdd-holm\endcsname{0}
\expandafter\def\csname pct@rev.pol.mdd.f20.uiv.p2.rcr-vs-full-recert-ni.mdd-holm\endcsname{1.4}
\expandafter\def\csname ppt@rev.pol.mdd.f20.uiv.p2.rcr-vs-full-recert-ni.mdd-holm\endcsname{$+$1.4}
\expandafter\def\csname val@rev.pol.mdd.f20.uiv.p2.rcr-vs-full-recert-ni.obs\endcsname{-0.002}
\expandafter\def\csname cnt@rev.pol.mdd.f20.uiv.p2.rcr-vs-full-recert-ni.obs\endcsname{0}
\expandafter\def\csname pct@rev.pol.mdd.f20.uiv.p2.rcr-vs-full-recert-ni.obs\endcsname{$-$0.2}
\expandafter\def\csname ppt@rev.pol.mdd.f20.uiv.p2.rcr-vs-full-recert-ni.obs\endcsname{$-$0.2}
\expandafter\def\csname val@rev.pol.mdd.f20.uiv.p2.rcr-vs-full-recert-ni.sd\endcsname{0.019}
\expandafter\def\csname cnt@rev.pol.mdd.f20.uiv.p2.rcr-vs-full-recert-ni.sd\endcsname{0}
\expandafter\def\csname pct@rev.pol.mdd.f20.uiv.p2.rcr-vs-full-recert-ni.sd\endcsname{1.9}
\expandafter\def\csname ppt@rev.pol.mdd.f20.uiv.p2.rcr-vs-full-recert-ni.sd\endcsname{$+$1.9}
\expandafter\def\csname val@rev.pol.mdd.f20.uiv.p2.rcr-vs-generic-ppi-source.G\endcsname{24}
\expandafter\def\csname cnt@rev.pol.mdd.f20.uiv.p2.rcr-vs-generic-ppi-source.G\endcsname{24}
\expandafter\def\csname pct@rev.pol.mdd.f20.uiv.p2.rcr-vs-generic-ppi-source.G\endcsname{2400.0}
\expandafter\def\csname ppt@rev.pol.mdd.f20.uiv.p2.rcr-vs-generic-ppi-source.G\endcsname{$+$2400.0}
\expandafter\def\csname val@rev.pol.mdd.f20.uiv.p2.rcr-vs-generic-ppi-source.mdd\endcsname{0.005}
\expandafter\def\csname cnt@rev.pol.mdd.f20.uiv.p2.rcr-vs-generic-ppi-source.mdd\endcsname{0}
\expandafter\def\csname pct@rev.pol.mdd.f20.uiv.p2.rcr-vs-generic-ppi-source.mdd\endcsname{0.5}
\expandafter\def\csname ppt@rev.pol.mdd.f20.uiv.p2.rcr-vs-generic-ppi-source.mdd\endcsname{$+$0.5}
\expandafter\def\csname val@rev.pol.mdd.f20.uiv.p2.rcr-vs-generic-ppi-source.mdd-holm\endcsname{0.006}
\expandafter\def\csname cnt@rev.pol.mdd.f20.uiv.p2.rcr-vs-generic-ppi-source.mdd-holm\endcsname{0}
\expandafter\def\csname pct@rev.pol.mdd.f20.uiv.p2.rcr-vs-generic-ppi-source.mdd-holm\endcsname{0.6}
\expandafter\def\csname ppt@rev.pol.mdd.f20.uiv.p2.rcr-vs-generic-ppi-source.mdd-holm\endcsname{$+$0.6}
\expandafter\def\csname val@rev.pol.mdd.f20.uiv.p2.rcr-vs-generic-ppi-source.obs\endcsname{0.004}
\expandafter\def\csname cnt@rev.pol.mdd.f20.uiv.p2.rcr-vs-generic-ppi-source.obs\endcsname{0}
\expandafter\def\csname pct@rev.pol.mdd.f20.uiv.p2.rcr-vs-generic-ppi-source.obs\endcsname{0.4}
\expandafter\def\csname ppt@rev.pol.mdd.f20.uiv.p2.rcr-vs-generic-ppi-source.obs\endcsname{$+$0.4}
\expandafter\def\csname val@rev.pol.mdd.f20.uiv.p2.rcr-vs-generic-ppi-source.sd\endcsname{0.009}
\expandafter\def\csname cnt@rev.pol.mdd.f20.uiv.p2.rcr-vs-generic-ppi-source.sd\endcsname{0}
\expandafter\def\csname pct@rev.pol.mdd.f20.uiv.p2.rcr-vs-generic-ppi-source.sd\endcsname{0.9}
\expandafter\def\csname ppt@rev.pol.mdd.f20.uiv.p2.rcr-vs-generic-ppi-source.sd\endcsname{$+$0.9}
\expandafter\def\csname val@rev.pol.mdd.f20.uiv.p2.rcr-vs-replay-ppi-uniform.G\endcsname{24}
\expandafter\def\csname cnt@rev.pol.mdd.f20.uiv.p2.rcr-vs-replay-ppi-uniform.G\endcsname{24}
\expandafter\def\csname pct@rev.pol.mdd.f20.uiv.p2.rcr-vs-replay-ppi-uniform.G\endcsname{2400.0}
\expandafter\def\csname ppt@rev.pol.mdd.f20.uiv.p2.rcr-vs-replay-ppi-uniform.G\endcsname{$+$2400.0}
\expandafter\def\csname val@rev.pol.mdd.f20.uiv.p2.rcr-vs-replay-ppi-uniform.mdd\endcsname{0.002}
\expandafter\def\csname cnt@rev.pol.mdd.f20.uiv.p2.rcr-vs-replay-ppi-uniform.mdd\endcsname{0}
\expandafter\def\csname pct@rev.pol.mdd.f20.uiv.p2.rcr-vs-replay-ppi-uniform.mdd\endcsname{0.2}
\expandafter\def\csname ppt@rev.pol.mdd.f20.uiv.p2.rcr-vs-replay-ppi-uniform.mdd\endcsname{$+$0.2}
\expandafter\def\csname val@rev.pol.mdd.f20.uiv.p2.rcr-vs-replay-ppi-uniform.mdd-holm\endcsname{0.003}
\expandafter\def\csname cnt@rev.pol.mdd.f20.uiv.p2.rcr-vs-replay-ppi-uniform.mdd-holm\endcsname{0}
\expandafter\def\csname pct@rev.pol.mdd.f20.uiv.p2.rcr-vs-replay-ppi-uniform.mdd-holm\endcsname{0.3}
\expandafter\def\csname ppt@rev.pol.mdd.f20.uiv.p2.rcr-vs-replay-ppi-uniform.mdd-holm\endcsname{$+$0.3}
\expandafter\def\csname val@rev.pol.mdd.f20.uiv.p2.rcr-vs-replay-ppi-uniform.obs\endcsname{0.002}
\expandafter\def\csname cnt@rev.pol.mdd.f20.uiv.p2.rcr-vs-replay-ppi-uniform.obs\endcsname{0}
\expandafter\def\csname pct@rev.pol.mdd.f20.uiv.p2.rcr-vs-replay-ppi-uniform.obs\endcsname{0.2}
\expandafter\def\csname ppt@rev.pol.mdd.f20.uiv.p2.rcr-vs-replay-ppi-uniform.obs\endcsname{$+$0.2}
\expandafter\def\csname val@rev.pol.mdd.f20.uiv.p2.rcr-vs-replay-ppi-uniform.sd\endcsname{0.004}
\expandafter\def\csname cnt@rev.pol.mdd.f20.uiv.p2.rcr-vs-replay-ppi-uniform.sd\endcsname{0}
\expandafter\def\csname pct@rev.pol.mdd.f20.uiv.p2.rcr-vs-replay-ppi-uniform.sd\endcsname{0.4}
\expandafter\def\csname ppt@rev.pol.mdd.f20.uiv.p2.rcr-vs-replay-ppi-uniform.sd\endcsname{$+$0.4}
\expandafter\def\csname val@rev.pol.mdd.n.sspro.p1.rcr-vs-full-recert-ni.G\endcsname{12}
\expandafter\def\csname cnt@rev.pol.mdd.n.sspro.p1.rcr-vs-full-recert-ni.G\endcsname{12}
\expandafter\def\csname pct@rev.pol.mdd.n.sspro.p1.rcr-vs-full-recert-ni.G\endcsname{1200.0}
\expandafter\def\csname ppt@rev.pol.mdd.n.sspro.p1.rcr-vs-full-recert-ni.G\endcsname{$+$1200.0}
\expandafter\def\csname val@rev.pol.mdd.n.sspro.p1.rcr-vs-full-recert-ni.mdd\endcsname{0.005}
\expandafter\def\csname cnt@rev.pol.mdd.n.sspro.p1.rcr-vs-full-recert-ni.mdd\endcsname{0}
\expandafter\def\csname pct@rev.pol.mdd.n.sspro.p1.rcr-vs-full-recert-ni.mdd\endcsname{0.5}
\expandafter\def\csname ppt@rev.pol.mdd.n.sspro.p1.rcr-vs-full-recert-ni.mdd\endcsname{$+$0.5}
\expandafter\def\csname val@rev.pol.mdd.n.sspro.p1.rcr-vs-full-recert-ni.mdd-holm\endcsname{0.008}
\expandafter\def\csname cnt@rev.pol.mdd.n.sspro.p1.rcr-vs-full-recert-ni.mdd-holm\endcsname{0}
\expandafter\def\csname pct@rev.pol.mdd.n.sspro.p1.rcr-vs-full-recert-ni.mdd-holm\endcsname{0.8}
\expandafter\def\csname ppt@rev.pol.mdd.n.sspro.p1.rcr-vs-full-recert-ni.mdd-holm\endcsname{$+$0.8}
\expandafter\def\csname val@rev.pol.mdd.n.sspro.p1.rcr-vs-full-recert-ni.obs\endcsname{-0.004}
\expandafter\def\csname cnt@rev.pol.mdd.n.sspro.p1.rcr-vs-full-recert-ni.obs\endcsname{0}
\expandafter\def\csname pct@rev.pol.mdd.n.sspro.p1.rcr-vs-full-recert-ni.obs\endcsname{$-$0.4}
\expandafter\def\csname ppt@rev.pol.mdd.n.sspro.p1.rcr-vs-full-recert-ni.obs\endcsname{$-$0.4}
\expandafter\def\csname val@rev.pol.mdd.n.sspro.p1.rcr-vs-full-recert-ni.sd\endcsname{0.007}
\expandafter\def\csname cnt@rev.pol.mdd.n.sspro.p1.rcr-vs-full-recert-ni.sd\endcsname{0}
\expandafter\def\csname pct@rev.pol.mdd.n.sspro.p1.rcr-vs-full-recert-ni.sd\endcsname{0.7}
\expandafter\def\csname ppt@rev.pol.mdd.n.sspro.p1.rcr-vs-full-recert-ni.sd\endcsname{$+$0.7}
\expandafter\def\csname val@rev.pol.mdd.n.sspro.p1.rcr-vs-generic-ppi-source.G\endcsname{12}
\expandafter\def\csname cnt@rev.pol.mdd.n.sspro.p1.rcr-vs-generic-ppi-source.G\endcsname{12}
\expandafter\def\csname pct@rev.pol.mdd.n.sspro.p1.rcr-vs-generic-ppi-source.G\endcsname{1200.0}
\expandafter\def\csname ppt@rev.pol.mdd.n.sspro.p1.rcr-vs-generic-ppi-source.G\endcsname{$+$1200.0}
\expandafter\def\csname val@rev.pol.mdd.n.sspro.p1.rcr-vs-generic-ppi-source.mdd\endcsname{0.012}
\expandafter\def\csname cnt@rev.pol.mdd.n.sspro.p1.rcr-vs-generic-ppi-source.mdd\endcsname{0}
\expandafter\def\csname pct@rev.pol.mdd.n.sspro.p1.rcr-vs-generic-ppi-source.mdd\endcsname{1.2}
\expandafter\def\csname ppt@rev.pol.mdd.n.sspro.p1.rcr-vs-generic-ppi-source.mdd\endcsname{$+$1.2}
\expandafter\def\csname val@rev.pol.mdd.n.sspro.p1.rcr-vs-generic-ppi-source.mdd-holm\endcsname{0.017}
\expandafter\def\csname cnt@rev.pol.mdd.n.sspro.p1.rcr-vs-generic-ppi-source.mdd-holm\endcsname{0}
\expandafter\def\csname pct@rev.pol.mdd.n.sspro.p1.rcr-vs-generic-ppi-source.mdd-holm\endcsname{1.7}
\expandafter\def\csname ppt@rev.pol.mdd.n.sspro.p1.rcr-vs-generic-ppi-source.mdd-holm\endcsname{$+$1.7}
\expandafter\def\csname val@rev.pol.mdd.n.sspro.p1.rcr-vs-generic-ppi-source.obs\endcsname{0.008}
\expandafter\def\csname cnt@rev.pol.mdd.n.sspro.p1.rcr-vs-generic-ppi-source.obs\endcsname{0}
\expandafter\def\csname pct@rev.pol.mdd.n.sspro.p1.rcr-vs-generic-ppi-source.obs\endcsname{0.8}
\expandafter\def\csname ppt@rev.pol.mdd.n.sspro.p1.rcr-vs-generic-ppi-source.obs\endcsname{$+$0.8}
\expandafter\def\csname val@rev.pol.mdd.n.sspro.p1.rcr-vs-generic-ppi-source.sd\endcsname{0.016}
\expandafter\def\csname cnt@rev.pol.mdd.n.sspro.p1.rcr-vs-generic-ppi-source.sd\endcsname{0}
\expandafter\def\csname pct@rev.pol.mdd.n.sspro.p1.rcr-vs-generic-ppi-source.sd\endcsname{1.6}
\expandafter\def\csname ppt@rev.pol.mdd.n.sspro.p1.rcr-vs-generic-ppi-source.sd\endcsname{$+$1.6}
\expandafter\def\csname val@rev.pol.mdd.n.sspro.p1.rcr-vs-replay-ppi-uniform.G\endcsname{12}
\expandafter\def\csname cnt@rev.pol.mdd.n.sspro.p1.rcr-vs-replay-ppi-uniform.G\endcsname{12}
\expandafter\def\csname pct@rev.pol.mdd.n.sspro.p1.rcr-vs-replay-ppi-uniform.G\endcsname{1200.0}
\expandafter\def\csname ppt@rev.pol.mdd.n.sspro.p1.rcr-vs-replay-ppi-uniform.G\endcsname{$+$1200.0}
\expandafter\def\csname val@rev.pol.mdd.n.sspro.p1.rcr-vs-replay-ppi-uniform.mdd\endcsname{0.005}
\expandafter\def\csname cnt@rev.pol.mdd.n.sspro.p1.rcr-vs-replay-ppi-uniform.mdd\endcsname{0}
\expandafter\def\csname pct@rev.pol.mdd.n.sspro.p1.rcr-vs-replay-ppi-uniform.mdd\endcsname{0.5}
\expandafter\def\csname ppt@rev.pol.mdd.n.sspro.p1.rcr-vs-replay-ppi-uniform.mdd\endcsname{$+$0.5}
\expandafter\def\csname val@rev.pol.mdd.n.sspro.p1.rcr-vs-replay-ppi-uniform.mdd-holm\endcsname{0.007}
\expandafter\def\csname cnt@rev.pol.mdd.n.sspro.p1.rcr-vs-replay-ppi-uniform.mdd-holm\endcsname{0}
\expandafter\def\csname pct@rev.pol.mdd.n.sspro.p1.rcr-vs-replay-ppi-uniform.mdd-holm\endcsname{0.7}
\expandafter\def\csname ppt@rev.pol.mdd.n.sspro.p1.rcr-vs-replay-ppi-uniform.mdd-holm\endcsname{$+$0.7}
\expandafter\def\csname val@rev.pol.mdd.n.sspro.p1.rcr-vs-replay-ppi-uniform.obs\endcsname{0.003}
\expandafter\def\csname cnt@rev.pol.mdd.n.sspro.p1.rcr-vs-replay-ppi-uniform.obs\endcsname{0}
\expandafter\def\csname pct@rev.pol.mdd.n.sspro.p1.rcr-vs-replay-ppi-uniform.obs\endcsname{0.3}
\expandafter\def\csname ppt@rev.pol.mdd.n.sspro.p1.rcr-vs-replay-ppi-uniform.obs\endcsname{$+$0.3}
\expandafter\def\csname val@rev.pol.mdd.n.sspro.p1.rcr-vs-replay-ppi-uniform.sd\endcsname{0.006}
\expandafter\def\csname cnt@rev.pol.mdd.n.sspro.p1.rcr-vs-replay-ppi-uniform.sd\endcsname{0}
\expandafter\def\csname pct@rev.pol.mdd.n.sspro.p1.rcr-vs-replay-ppi-uniform.sd\endcsname{0.6}
\expandafter\def\csname ppt@rev.pol.mdd.n.sspro.p1.rcr-vs-replay-ppi-uniform.sd\endcsname{$+$0.6}
\expandafter\def\csname val@rev.pol.mdd.n.sspro.p2.rcr-vs-full-recert-ni.G\endcsname{12}
\expandafter\def\csname cnt@rev.pol.mdd.n.sspro.p2.rcr-vs-full-recert-ni.G\endcsname{12}
\expandafter\def\csname pct@rev.pol.mdd.n.sspro.p2.rcr-vs-full-recert-ni.G\endcsname{1200.0}
\expandafter\def\csname ppt@rev.pol.mdd.n.sspro.p2.rcr-vs-full-recert-ni.G\endcsname{$+$1200.0}
\expandafter\def\csname val@rev.pol.mdd.n.sspro.p2.rcr-vs-full-recert-ni.mdd\endcsname{0.010}
\expandafter\def\csname cnt@rev.pol.mdd.n.sspro.p2.rcr-vs-full-recert-ni.mdd\endcsname{0}
\expandafter\def\csname pct@rev.pol.mdd.n.sspro.p2.rcr-vs-full-recert-ni.mdd\endcsname{1.0}
\expandafter\def\csname ppt@rev.pol.mdd.n.sspro.p2.rcr-vs-full-recert-ni.mdd\endcsname{$+$1.0}
\expandafter\def\csname val@rev.pol.mdd.n.sspro.p2.rcr-vs-full-recert-ni.mdd-holm\endcsname{0.015}
\expandafter\def\csname cnt@rev.pol.mdd.n.sspro.p2.rcr-vs-full-recert-ni.mdd-holm\endcsname{0}
\expandafter\def\csname pct@rev.pol.mdd.n.sspro.p2.rcr-vs-full-recert-ni.mdd-holm\endcsname{1.5}
\expandafter\def\csname ppt@rev.pol.mdd.n.sspro.p2.rcr-vs-full-recert-ni.mdd-holm\endcsname{$+$1.5}
\expandafter\def\csname val@rev.pol.mdd.n.sspro.p2.rcr-vs-full-recert-ni.obs\endcsname{-0.016}
\expandafter\def\csname cnt@rev.pol.mdd.n.sspro.p2.rcr-vs-full-recert-ni.obs\endcsname{0}
\expandafter\def\csname pct@rev.pol.mdd.n.sspro.p2.rcr-vs-full-recert-ni.obs\endcsname{$-$1.6}
\expandafter\def\csname ppt@rev.pol.mdd.n.sspro.p2.rcr-vs-full-recert-ni.obs\endcsname{$-$1.6}
\expandafter\def\csname val@rev.pol.mdd.n.sspro.p2.rcr-vs-full-recert-ni.sd\endcsname{0.014}
\expandafter\def\csname cnt@rev.pol.mdd.n.sspro.p2.rcr-vs-full-recert-ni.sd\endcsname{0}
\expandafter\def\csname pct@rev.pol.mdd.n.sspro.p2.rcr-vs-full-recert-ni.sd\endcsname{1.4}
\expandafter\def\csname ppt@rev.pol.mdd.n.sspro.p2.rcr-vs-full-recert-ni.sd\endcsname{$+$1.4}
\expandafter\def\csname val@rev.pol.mdd.n.sspro.p2.rcr-vs-generic-ppi-source.G\endcsname{12}
\expandafter\def\csname cnt@rev.pol.mdd.n.sspro.p2.rcr-vs-generic-ppi-source.G\endcsname{12}
\expandafter\def\csname pct@rev.pol.mdd.n.sspro.p2.rcr-vs-generic-ppi-source.G\endcsname{1200.0}
\expandafter\def\csname ppt@rev.pol.mdd.n.sspro.p2.rcr-vs-generic-ppi-source.G\endcsname{$+$1200.0}
\expandafter\def\csname val@rev.pol.mdd.n.sspro.p2.rcr-vs-generic-ppi-source.mdd\endcsname{0.001}
\expandafter\def\csname cnt@rev.pol.mdd.n.sspro.p2.rcr-vs-generic-ppi-source.mdd\endcsname{0}
\expandafter\def\csname pct@rev.pol.mdd.n.sspro.p2.rcr-vs-generic-ppi-source.mdd\endcsname{0.1}
\expandafter\def\csname ppt@rev.pol.mdd.n.sspro.p2.rcr-vs-generic-ppi-source.mdd\endcsname{$+$0.1}
\expandafter\def\csname val@rev.pol.mdd.n.sspro.p2.rcr-vs-generic-ppi-source.mdd-holm\endcsname{0.001}
\expandafter\def\csname cnt@rev.pol.mdd.n.sspro.p2.rcr-vs-generic-ppi-source.mdd-holm\endcsname{0}
\expandafter\def\csname pct@rev.pol.mdd.n.sspro.p2.rcr-vs-generic-ppi-source.mdd-holm\endcsname{0.1}
\expandafter\def\csname ppt@rev.pol.mdd.n.sspro.p2.rcr-vs-generic-ppi-source.mdd-holm\endcsname{$+$0.1}
\expandafter\def\csname val@rev.pol.mdd.n.sspro.p2.rcr-vs-generic-ppi-source.obs\endcsname{-0.000}
\expandafter\def\csname cnt@rev.pol.mdd.n.sspro.p2.rcr-vs-generic-ppi-source.obs\endcsname{0}
\expandafter\def\csname pct@rev.pol.mdd.n.sspro.p2.rcr-vs-generic-ppi-source.obs\endcsname{0.0}
\expandafter\def\csname ppt@rev.pol.mdd.n.sspro.p2.rcr-vs-generic-ppi-source.obs\endcsname{0.0}
\expandafter\def\csname val@rev.pol.mdd.n.sspro.p2.rcr-vs-generic-ppi-source.sd\endcsname{0.001}
\expandafter\def\csname cnt@rev.pol.mdd.n.sspro.p2.rcr-vs-generic-ppi-source.sd\endcsname{0}
\expandafter\def\csname pct@rev.pol.mdd.n.sspro.p2.rcr-vs-generic-ppi-source.sd\endcsname{0.1}
\expandafter\def\csname ppt@rev.pol.mdd.n.sspro.p2.rcr-vs-generic-ppi-source.sd\endcsname{$+$0.1}
\expandafter\def\csname val@rev.pol.mdd.n.sspro.p2.rcr-vs-replay-ppi-uniform.G\endcsname{12}
\expandafter\def\csname cnt@rev.pol.mdd.n.sspro.p2.rcr-vs-replay-ppi-uniform.G\endcsname{12}
\expandafter\def\csname pct@rev.pol.mdd.n.sspro.p2.rcr-vs-replay-ppi-uniform.G\endcsname{1200.0}
\expandafter\def\csname ppt@rev.pol.mdd.n.sspro.p2.rcr-vs-replay-ppi-uniform.G\endcsname{$+$1200.0}
\expandafter\def\csname val@rev.pol.mdd.n.sspro.p2.rcr-vs-replay-ppi-uniform.mdd\endcsname{0.001}
\expandafter\def\csname cnt@rev.pol.mdd.n.sspro.p2.rcr-vs-replay-ppi-uniform.mdd\endcsname{0}
\expandafter\def\csname pct@rev.pol.mdd.n.sspro.p2.rcr-vs-replay-ppi-uniform.mdd\endcsname{0.1}
\expandafter\def\csname ppt@rev.pol.mdd.n.sspro.p2.rcr-vs-replay-ppi-uniform.mdd\endcsname{$+$0.1}
\expandafter\def\csname val@rev.pol.mdd.n.sspro.p2.rcr-vs-replay-ppi-uniform.mdd-holm\endcsname{0.001}
\expandafter\def\csname cnt@rev.pol.mdd.n.sspro.p2.rcr-vs-replay-ppi-uniform.mdd-holm\endcsname{0}
\expandafter\def\csname pct@rev.pol.mdd.n.sspro.p2.rcr-vs-replay-ppi-uniform.mdd-holm\endcsname{0.1}
\expandafter\def\csname ppt@rev.pol.mdd.n.sspro.p2.rcr-vs-replay-ppi-uniform.mdd-holm\endcsname{$+$0.1}
\expandafter\def\csname val@rev.pol.mdd.n.sspro.p2.rcr-vs-replay-ppi-uniform.obs\endcsname{0.000}
\expandafter\def\csname cnt@rev.pol.mdd.n.sspro.p2.rcr-vs-replay-ppi-uniform.obs\endcsname{0}
\expandafter\def\csname pct@rev.pol.mdd.n.sspro.p2.rcr-vs-replay-ppi-uniform.obs\endcsname{0.0}
\expandafter\def\csname ppt@rev.pol.mdd.n.sspro.p2.rcr-vs-replay-ppi-uniform.obs\endcsname{0.0}
\expandafter\def\csname val@rev.pol.mdd.n.sspro.p2.rcr-vs-replay-ppi-uniform.sd\endcsname{0.001}
\expandafter\def\csname cnt@rev.pol.mdd.n.sspro.p2.rcr-vs-replay-ppi-uniform.sd\endcsname{0}
\expandafter\def\csname pct@rev.pol.mdd.n.sspro.p2.rcr-vs-replay-ppi-uniform.sd\endcsname{0.1}
\expandafter\def\csname ppt@rev.pol.mdd.n.sspro.p2.rcr-vs-replay-ppi-uniform.sd\endcsname{$+$0.1}
\expandafter\def\csname val@rev.pol.mdd.n.uiv.p1.rcr-vs-full-recert-ni.G\endcsname{24}
\expandafter\def\csname cnt@rev.pol.mdd.n.uiv.p1.rcr-vs-full-recert-ni.G\endcsname{24}
\expandafter\def\csname pct@rev.pol.mdd.n.uiv.p1.rcr-vs-full-recert-ni.G\endcsname{2400.0}
\expandafter\def\csname ppt@rev.pol.mdd.n.uiv.p1.rcr-vs-full-recert-ni.G\endcsname{$+$2400.0}
\expandafter\def\csname val@rev.pol.mdd.n.uiv.p1.rcr-vs-full-recert-ni.mdd\endcsname{0.012}
\expandafter\def\csname cnt@rev.pol.mdd.n.uiv.p1.rcr-vs-full-recert-ni.mdd\endcsname{0}
\expandafter\def\csname pct@rev.pol.mdd.n.uiv.p1.rcr-vs-full-recert-ni.mdd\endcsname{1.2}
\expandafter\def\csname ppt@rev.pol.mdd.n.uiv.p1.rcr-vs-full-recert-ni.mdd\endcsname{$+$1.2}
\expandafter\def\csname val@rev.pol.mdd.n.uiv.p1.rcr-vs-full-recert-ni.mdd-holm\endcsname{0.017}
\expandafter\def\csname cnt@rev.pol.mdd.n.uiv.p1.rcr-vs-full-recert-ni.mdd-holm\endcsname{0}
\expandafter\def\csname pct@rev.pol.mdd.n.uiv.p1.rcr-vs-full-recert-ni.mdd-holm\endcsname{1.7}
\expandafter\def\csname ppt@rev.pol.mdd.n.uiv.p1.rcr-vs-full-recert-ni.mdd-holm\endcsname{$+$1.7}
\expandafter\def\csname val@rev.pol.mdd.n.uiv.p1.rcr-vs-full-recert-ni.obs\endcsname{0.011}
\expandafter\def\csname cnt@rev.pol.mdd.n.uiv.p1.rcr-vs-full-recert-ni.obs\endcsname{0}
\expandafter\def\csname pct@rev.pol.mdd.n.uiv.p1.rcr-vs-full-recert-ni.obs\endcsname{1.1}
\expandafter\def\csname ppt@rev.pol.mdd.n.uiv.p1.rcr-vs-full-recert-ni.obs\endcsname{$+$1.1}
\expandafter\def\csname val@rev.pol.mdd.n.uiv.p1.rcr-vs-full-recert-ni.sd\endcsname{0.024}
\expandafter\def\csname cnt@rev.pol.mdd.n.uiv.p1.rcr-vs-full-recert-ni.sd\endcsname{0}
\expandafter\def\csname pct@rev.pol.mdd.n.uiv.p1.rcr-vs-full-recert-ni.sd\endcsname{2.4}
\expandafter\def\csname ppt@rev.pol.mdd.n.uiv.p1.rcr-vs-full-recert-ni.sd\endcsname{$+$2.4}
\expandafter\def\csname val@rev.pol.mdd.n.uiv.p1.rcr-vs-generic-ppi-source.G\endcsname{24}
\expandafter\def\csname cnt@rev.pol.mdd.n.uiv.p1.rcr-vs-generic-ppi-source.G\endcsname{24}
\expandafter\def\csname pct@rev.pol.mdd.n.uiv.p1.rcr-vs-generic-ppi-source.G\endcsname{2400.0}
\expandafter\def\csname ppt@rev.pol.mdd.n.uiv.p1.rcr-vs-generic-ppi-source.G\endcsname{$+$2400.0}
\expandafter\def\csname val@rev.pol.mdd.n.uiv.p1.rcr-vs-generic-ppi-source.mdd\endcsname{0.006}
\expandafter\def\csname cnt@rev.pol.mdd.n.uiv.p1.rcr-vs-generic-ppi-source.mdd\endcsname{0}
\expandafter\def\csname pct@rev.pol.mdd.n.uiv.p1.rcr-vs-generic-ppi-source.mdd\endcsname{0.6}
\expandafter\def\csname ppt@rev.pol.mdd.n.uiv.p1.rcr-vs-generic-ppi-source.mdd\endcsname{$+$0.6}
\expandafter\def\csname val@rev.pol.mdd.n.uiv.p1.rcr-vs-generic-ppi-source.mdd-holm\endcsname{0.008}
\expandafter\def\csname cnt@rev.pol.mdd.n.uiv.p1.rcr-vs-generic-ppi-source.mdd-holm\endcsname{0}
\expandafter\def\csname pct@rev.pol.mdd.n.uiv.p1.rcr-vs-generic-ppi-source.mdd-holm\endcsname{0.8}
\expandafter\def\csname ppt@rev.pol.mdd.n.uiv.p1.rcr-vs-generic-ppi-source.mdd-holm\endcsname{$+$0.8}
\expandafter\def\csname val@rev.pol.mdd.n.uiv.p1.rcr-vs-generic-ppi-source.obs\endcsname{0.008}
\expandafter\def\csname cnt@rev.pol.mdd.n.uiv.p1.rcr-vs-generic-ppi-source.obs\endcsname{0}
\expandafter\def\csname pct@rev.pol.mdd.n.uiv.p1.rcr-vs-generic-ppi-source.obs\endcsname{0.8}
\expandafter\def\csname ppt@rev.pol.mdd.n.uiv.p1.rcr-vs-generic-ppi-source.obs\endcsname{$+$0.8}
\expandafter\def\csname val@rev.pol.mdd.n.uiv.p1.rcr-vs-generic-ppi-source.sd\endcsname{0.011}
\expandafter\def\csname cnt@rev.pol.mdd.n.uiv.p1.rcr-vs-generic-ppi-source.sd\endcsname{0}
\expandafter\def\csname pct@rev.pol.mdd.n.uiv.p1.rcr-vs-generic-ppi-source.sd\endcsname{1.1}
\expandafter\def\csname ppt@rev.pol.mdd.n.uiv.p1.rcr-vs-generic-ppi-source.sd\endcsname{$+$1.1}
\expandafter\def\csname val@rev.pol.mdd.n.uiv.p1.rcr-vs-replay-ppi-uniform.G\endcsname{24}
\expandafter\def\csname cnt@rev.pol.mdd.n.uiv.p1.rcr-vs-replay-ppi-uniform.G\endcsname{24}
\expandafter\def\csname pct@rev.pol.mdd.n.uiv.p1.rcr-vs-replay-ppi-uniform.G\endcsname{2400.0}
\expandafter\def\csname ppt@rev.pol.mdd.n.uiv.p1.rcr-vs-replay-ppi-uniform.G\endcsname{$+$2400.0}
\expandafter\def\csname val@rev.pol.mdd.n.uiv.p1.rcr-vs-replay-ppi-uniform.mdd\endcsname{0.003}
\expandafter\def\csname cnt@rev.pol.mdd.n.uiv.p1.rcr-vs-replay-ppi-uniform.mdd\endcsname{0}
\expandafter\def\csname pct@rev.pol.mdd.n.uiv.p1.rcr-vs-replay-ppi-uniform.mdd\endcsname{0.3}
\expandafter\def\csname ppt@rev.pol.mdd.n.uiv.p1.rcr-vs-replay-ppi-uniform.mdd\endcsname{$+$0.3}
\expandafter\def\csname val@rev.pol.mdd.n.uiv.p1.rcr-vs-replay-ppi-uniform.mdd-holm\endcsname{0.004}
\expandafter\def\csname cnt@rev.pol.mdd.n.uiv.p1.rcr-vs-replay-ppi-uniform.mdd-holm\endcsname{0}
\expandafter\def\csname pct@rev.pol.mdd.n.uiv.p1.rcr-vs-replay-ppi-uniform.mdd-holm\endcsname{0.4}
\expandafter\def\csname ppt@rev.pol.mdd.n.uiv.p1.rcr-vs-replay-ppi-uniform.mdd-holm\endcsname{$+$0.4}
\expandafter\def\csname val@rev.pol.mdd.n.uiv.p1.rcr-vs-replay-ppi-uniform.obs\endcsname{0.004}
\expandafter\def\csname cnt@rev.pol.mdd.n.uiv.p1.rcr-vs-replay-ppi-uniform.obs\endcsname{0}
\expandafter\def\csname pct@rev.pol.mdd.n.uiv.p1.rcr-vs-replay-ppi-uniform.obs\endcsname{0.4}
\expandafter\def\csname ppt@rev.pol.mdd.n.uiv.p1.rcr-vs-replay-ppi-uniform.obs\endcsname{$+$0.4}
\expandafter\def\csname val@rev.pol.mdd.n.uiv.p1.rcr-vs-replay-ppi-uniform.sd\endcsname{0.006}
\expandafter\def\csname cnt@rev.pol.mdd.n.uiv.p1.rcr-vs-replay-ppi-uniform.sd\endcsname{0}
\expandafter\def\csname pct@rev.pol.mdd.n.uiv.p1.rcr-vs-replay-ppi-uniform.sd\endcsname{0.6}
\expandafter\def\csname ppt@rev.pol.mdd.n.uiv.p1.rcr-vs-replay-ppi-uniform.sd\endcsname{$+$0.6}
\expandafter\def\csname val@rev.pol.mdd.n.uiv.p2.rcr-vs-full-recert-ni.G\endcsname{24}
\expandafter\def\csname cnt@rev.pol.mdd.n.uiv.p2.rcr-vs-full-recert-ni.G\endcsname{24}
\expandafter\def\csname pct@rev.pol.mdd.n.uiv.p2.rcr-vs-full-recert-ni.G\endcsname{2400.0}
\expandafter\def\csname ppt@rev.pol.mdd.n.uiv.p2.rcr-vs-full-recert-ni.G\endcsname{$+$2400.0}
\expandafter\def\csname val@rev.pol.mdd.n.uiv.p2.rcr-vs-full-recert-ni.mdd\endcsname{0.007}
\expandafter\def\csname cnt@rev.pol.mdd.n.uiv.p2.rcr-vs-full-recert-ni.mdd\endcsname{0}
\expandafter\def\csname pct@rev.pol.mdd.n.uiv.p2.rcr-vs-full-recert-ni.mdd\endcsname{0.7}
\expandafter\def\csname ppt@rev.pol.mdd.n.uiv.p2.rcr-vs-full-recert-ni.mdd\endcsname{$+$0.7}
\expandafter\def\csname val@rev.pol.mdd.n.uiv.p2.rcr-vs-full-recert-ni.mdd-holm\endcsname{0.010}
\expandafter\def\csname cnt@rev.pol.mdd.n.uiv.p2.rcr-vs-full-recert-ni.mdd-holm\endcsname{0}
\expandafter\def\csname pct@rev.pol.mdd.n.uiv.p2.rcr-vs-full-recert-ni.mdd-holm\endcsname{1.0}
\expandafter\def\csname ppt@rev.pol.mdd.n.uiv.p2.rcr-vs-full-recert-ni.mdd-holm\endcsname{$+$1.0}
\expandafter\def\csname val@rev.pol.mdd.n.uiv.p2.rcr-vs-full-recert-ni.obs\endcsname{-0.003}
\expandafter\def\csname cnt@rev.pol.mdd.n.uiv.p2.rcr-vs-full-recert-ni.obs\endcsname{0}
\expandafter\def\csname pct@rev.pol.mdd.n.uiv.p2.rcr-vs-full-recert-ni.obs\endcsname{$-$0.3}
\expandafter\def\csname ppt@rev.pol.mdd.n.uiv.p2.rcr-vs-full-recert-ni.obs\endcsname{$-$0.3}
\expandafter\def\csname val@rev.pol.mdd.n.uiv.p2.rcr-vs-full-recert-ni.sd\endcsname{0.014}
\expandafter\def\csname cnt@rev.pol.mdd.n.uiv.p2.rcr-vs-full-recert-ni.sd\endcsname{0}
\expandafter\def\csname pct@rev.pol.mdd.n.uiv.p2.rcr-vs-full-recert-ni.sd\endcsname{1.4}
\expandafter\def\csname ppt@rev.pol.mdd.n.uiv.p2.rcr-vs-full-recert-ni.sd\endcsname{$+$1.4}
\expandafter\def\csname val@rev.pol.mdd.n.uiv.p2.rcr-vs-generic-ppi-source.G\endcsname{24}
\expandafter\def\csname cnt@rev.pol.mdd.n.uiv.p2.rcr-vs-generic-ppi-source.G\endcsname{24}
\expandafter\def\csname pct@rev.pol.mdd.n.uiv.p2.rcr-vs-generic-ppi-source.G\endcsname{2400.0}
\expandafter\def\csname ppt@rev.pol.mdd.n.uiv.p2.rcr-vs-generic-ppi-source.G\endcsname{$+$2400.0}
\expandafter\def\csname val@rev.pol.mdd.n.uiv.p2.rcr-vs-generic-ppi-source.mdd\endcsname{0.002}
\expandafter\def\csname cnt@rev.pol.mdd.n.uiv.p2.rcr-vs-generic-ppi-source.mdd\endcsname{0}
\expandafter\def\csname pct@rev.pol.mdd.n.uiv.p2.rcr-vs-generic-ppi-source.mdd\endcsname{0.2}
\expandafter\def\csname ppt@rev.pol.mdd.n.uiv.p2.rcr-vs-generic-ppi-source.mdd\endcsname{$+$0.2}
\expandafter\def\csname val@rev.pol.mdd.n.uiv.p2.rcr-vs-generic-ppi-source.mdd-holm\endcsname{0.003}
\expandafter\def\csname cnt@rev.pol.mdd.n.uiv.p2.rcr-vs-generic-ppi-source.mdd-holm\endcsname{0}
\expandafter\def\csname pct@rev.pol.mdd.n.uiv.p2.rcr-vs-generic-ppi-source.mdd-holm\endcsname{0.3}
\expandafter\def\csname ppt@rev.pol.mdd.n.uiv.p2.rcr-vs-generic-ppi-source.mdd-holm\endcsname{$+$0.3}
\expandafter\def\csname val@rev.pol.mdd.n.uiv.p2.rcr-vs-generic-ppi-source.obs\endcsname{0.001}
\expandafter\def\csname cnt@rev.pol.mdd.n.uiv.p2.rcr-vs-generic-ppi-source.obs\endcsname{0}
\expandafter\def\csname pct@rev.pol.mdd.n.uiv.p2.rcr-vs-generic-ppi-source.obs\endcsname{0.1}
\expandafter\def\csname ppt@rev.pol.mdd.n.uiv.p2.rcr-vs-generic-ppi-source.obs\endcsname{$+$0.1}
\expandafter\def\csname val@rev.pol.mdd.n.uiv.p2.rcr-vs-generic-ppi-source.sd\endcsname{0.004}
\expandafter\def\csname cnt@rev.pol.mdd.n.uiv.p2.rcr-vs-generic-ppi-source.sd\endcsname{0}
\expandafter\def\csname pct@rev.pol.mdd.n.uiv.p2.rcr-vs-generic-ppi-source.sd\endcsname{0.4}
\expandafter\def\csname ppt@rev.pol.mdd.n.uiv.p2.rcr-vs-generic-ppi-source.sd\endcsname{$+$0.4}
\expandafter\def\csname val@rev.pol.mdd.n.uiv.p2.rcr-vs-replay-ppi-uniform.G\endcsname{24}
\expandafter\def\csname cnt@rev.pol.mdd.n.uiv.p2.rcr-vs-replay-ppi-uniform.G\endcsname{24}
\expandafter\def\csname pct@rev.pol.mdd.n.uiv.p2.rcr-vs-replay-ppi-uniform.G\endcsname{2400.0}
\expandafter\def\csname ppt@rev.pol.mdd.n.uiv.p2.rcr-vs-replay-ppi-uniform.G\endcsname{$+$2400.0}
\expandafter\def\csname val@rev.pol.mdd.n.uiv.p2.rcr-vs-replay-ppi-uniform.mdd\endcsname{0.001}
\expandafter\def\csname cnt@rev.pol.mdd.n.uiv.p2.rcr-vs-replay-ppi-uniform.mdd\endcsname{0}
\expandafter\def\csname pct@rev.pol.mdd.n.uiv.p2.rcr-vs-replay-ppi-uniform.mdd\endcsname{0.1}
\expandafter\def\csname ppt@rev.pol.mdd.n.uiv.p2.rcr-vs-replay-ppi-uniform.mdd\endcsname{$+$0.1}
\expandafter\def\csname val@rev.pol.mdd.n.uiv.p2.rcr-vs-replay-ppi-uniform.mdd-holm\endcsname{0.001}
\expandafter\def\csname cnt@rev.pol.mdd.n.uiv.p2.rcr-vs-replay-ppi-uniform.mdd-holm\endcsname{0}
\expandafter\def\csname pct@rev.pol.mdd.n.uiv.p2.rcr-vs-replay-ppi-uniform.mdd-holm\endcsname{0.1}
\expandafter\def\csname ppt@rev.pol.mdd.n.uiv.p2.rcr-vs-replay-ppi-uniform.mdd-holm\endcsname{$+$0.1}
\expandafter\def\csname val@rev.pol.mdd.n.uiv.p2.rcr-vs-replay-ppi-uniform.obs\endcsname{0.001}
\expandafter\def\csname cnt@rev.pol.mdd.n.uiv.p2.rcr-vs-replay-ppi-uniform.obs\endcsname{0}
\expandafter\def\csname pct@rev.pol.mdd.n.uiv.p2.rcr-vs-replay-ppi-uniform.obs\endcsname{0.1}
\expandafter\def\csname ppt@rev.pol.mdd.n.uiv.p2.rcr-vs-replay-ppi-uniform.obs\endcsname{$+$0.1}
\expandafter\def\csname val@rev.pol.mdd.n.uiv.p2.rcr-vs-replay-ppi-uniform.sd\endcsname{0.002}
\expandafter\def\csname cnt@rev.pol.mdd.n.uiv.p2.rcr-vs-replay-ppi-uniform.sd\endcsname{0}
\expandafter\def\csname pct@rev.pol.mdd.n.uiv.p2.rcr-vs-replay-ppi-uniform.sd\endcsname{0.2}
\expandafter\def\csname ppt@rev.pol.mdd.n.uiv.p2.rcr-vs-replay-ppi-uniform.sd\endcsname{$+$0.2}
\expandafter\def\csname val@rev.pol.n-hist\endcsname{10000}
\expandafter\def\csname cnt@rev.pol.n-hist\endcsname{10{,}000}
\expandafter\def\csname pct@rev.pol.n-hist\endcsname{1000000.0}
\expandafter\def\csname ppt@rev.pol.n-hist\endcsname{$+$1000000.0}
\expandafter\def\csname val@rev.pol.n-hist-frozen\endcsname{20}
\expandafter\def\csname cnt@rev.pol.n-hist-frozen\endcsname{20}
\expandafter\def\csname pct@rev.pol.n-hist-frozen\endcsname{2000.0}
\expandafter\def\csname ppt@rev.pol.n-hist-frozen\endcsname{$+$2000.0}
\expandafter\def\csname val@rev.pol.ni.f20.n-pass\endcsname{3}
\expandafter\def\csname cnt@rev.pol.ni.f20.n-pass\endcsname{3}
\expandafter\def\csname pct@rev.pol.ni.f20.n-pass\endcsname{300.0}
\expandafter\def\csname ppt@rev.pol.ni.f20.n-pass\endcsname{$+$300.0}
\expandafter\def\csname val@rev.pol.ni.f20.sspro.p1.lb\endcsname{-0.003}
\expandafter\def\csname cnt@rev.pol.ni.f20.sspro.p1.lb\endcsname{0}
\expandafter\def\csname pct@rev.pol.ni.f20.sspro.p1.lb\endcsname{$-$0.3}
\expandafter\def\csname ppt@rev.pol.ni.f20.sspro.p1.lb\endcsname{$-$0.3}
\expandafter\def\csname val@rev.pol.ni.f20.sspro.p1.mean\endcsname{-0.002}
\expandafter\def\csname cnt@rev.pol.ni.f20.sspro.p1.mean\endcsname{0}
\expandafter\def\csname pct@rev.pol.ni.f20.sspro.p1.mean\endcsname{$-$0.2}
\expandafter\def\csname ppt@rev.pol.ni.f20.sspro.p1.mean\endcsname{$-$0.2}
\expandafter\def\csname val@rev.pol.ni.f20.sspro.p1.pass\endcsname{1}
\expandafter\def\csname cnt@rev.pol.ni.f20.sspro.p1.pass\endcsname{1}
\expandafter\def\csname pct@rev.pol.ni.f20.sspro.p1.pass\endcsname{100.0}
\expandafter\def\csname ppt@rev.pol.ni.f20.sspro.p1.pass\endcsname{$+$100.0}
\expandafter\def\csname val@rev.pol.ni.f20.sspro.p2.lb\endcsname{-0.016}
\expandafter\def\csname cnt@rev.pol.ni.f20.sspro.p2.lb\endcsname{0}
\expandafter\def\csname pct@rev.pol.ni.f20.sspro.p2.lb\endcsname{$-$1.6}
\expandafter\def\csname ppt@rev.pol.ni.f20.sspro.p2.lb\endcsname{$-$1.6}
\expandafter\def\csname val@rev.pol.ni.f20.sspro.p2.mean\endcsname{-0.011}
\expandafter\def\csname cnt@rev.pol.ni.f20.sspro.p2.mean\endcsname{0}
\expandafter\def\csname pct@rev.pol.ni.f20.sspro.p2.mean\endcsname{$-$1.1}
\expandafter\def\csname ppt@rev.pol.ni.f20.sspro.p2.mean\endcsname{$-$1.1}
\expandafter\def\csname val@rev.pol.ni.f20.sspro.p2.pass\endcsname{0}
\expandafter\def\csname cnt@rev.pol.ni.f20.sspro.p2.pass\endcsname{0}
\expandafter\def\csname pct@rev.pol.ni.f20.sspro.p2.pass\endcsname{0.0}
\expandafter\def\csname ppt@rev.pol.ni.f20.sspro.p2.pass\endcsname{0.0}
\expandafter\def\csname val@rev.pol.ni.f20.uiv.p1.lb\endcsname{0.003}
\expandafter\def\csname cnt@rev.pol.ni.f20.uiv.p1.lb\endcsname{0}
\expandafter\def\csname pct@rev.pol.ni.f20.uiv.p1.lb\endcsname{0.3}
\expandafter\def\csname ppt@rev.pol.ni.f20.uiv.p1.lb\endcsname{$+$0.3}
\expandafter\def\csname val@rev.pol.ni.f20.uiv.p1.mean\endcsname{0.010}
\expandafter\def\csname cnt@rev.pol.ni.f20.uiv.p1.mean\endcsname{0}
\expandafter\def\csname pct@rev.pol.ni.f20.uiv.p1.mean\endcsname{1.0}
\expandafter\def\csname ppt@rev.pol.ni.f20.uiv.p1.mean\endcsname{$+$1.0}
\expandafter\def\csname val@rev.pol.ni.f20.uiv.p1.pass\endcsname{1}
\expandafter\def\csname cnt@rev.pol.ni.f20.uiv.p1.pass\endcsname{1}
\expandafter\def\csname pct@rev.pol.ni.f20.uiv.p1.pass\endcsname{100.0}
\expandafter\def\csname ppt@rev.pol.ni.f20.uiv.p1.pass\endcsname{$+$100.0}
\expandafter\def\csname val@rev.pol.ni.f20.uiv.p2.lb\endcsname{-0.008}
\expandafter\def\csname cnt@rev.pol.ni.f20.uiv.p2.lb\endcsname{0}
\expandafter\def\csname pct@rev.pol.ni.f20.uiv.p2.lb\endcsname{$-$0.8}
\expandafter\def\csname ppt@rev.pol.ni.f20.uiv.p2.lb\endcsname{$-$0.8}
\expandafter\def\csname val@rev.pol.ni.f20.uiv.p2.mean\endcsname{-0.002}
\expandafter\def\csname cnt@rev.pol.ni.f20.uiv.p2.mean\endcsname{0}
\expandafter\def\csname pct@rev.pol.ni.f20.uiv.p2.mean\endcsname{$-$0.2}
\expandafter\def\csname ppt@rev.pol.ni.f20.uiv.p2.mean\endcsname{$-$0.2}
\expandafter\def\csname val@rev.pol.ni.f20.uiv.p2.pass\endcsname{1}
\expandafter\def\csname cnt@rev.pol.ni.f20.uiv.p2.pass\endcsname{1}
\expandafter\def\csname pct@rev.pol.ni.f20.uiv.p2.pass\endcsname{100.0}
\expandafter\def\csname ppt@rev.pol.ni.f20.uiv.p2.pass\endcsname{$+$100.0}
\expandafter\def\csname val@rev.pol.ni.n.n-pass\endcsname{3}
\expandafter\def\csname cnt@rev.pol.ni.n.n-pass\endcsname{3}
\expandafter\def\csname pct@rev.pol.ni.n.n-pass\endcsname{300.0}
\expandafter\def\csname ppt@rev.pol.ni.n.n-pass\endcsname{$+$300.0}
\expandafter\def\csname val@rev.pol.ni.n.sspro.p1.lb\endcsname{-0.007}
\expandafter\def\csname cnt@rev.pol.ni.n.sspro.p1.lb\endcsname{0}
\expandafter\def\csname pct@rev.pol.ni.n.sspro.p1.lb\endcsname{$-$0.7}
\expandafter\def\csname ppt@rev.pol.ni.n.sspro.p1.lb\endcsname{$-$0.7}
\expandafter\def\csname val@rev.pol.ni.n.sspro.p1.mean\endcsname{-0.004}
\expandafter\def\csname cnt@rev.pol.ni.n.sspro.p1.mean\endcsname{0}
\expandafter\def\csname pct@rev.pol.ni.n.sspro.p1.mean\endcsname{$-$0.4}
\expandafter\def\csname ppt@rev.pol.ni.n.sspro.p1.mean\endcsname{$-$0.4}
\expandafter\def\csname val@rev.pol.ni.n.sspro.p1.pass\endcsname{1}
\expandafter\def\csname cnt@rev.pol.ni.n.sspro.p1.pass\endcsname{1}
\expandafter\def\csname pct@rev.pol.ni.n.sspro.p1.pass\endcsname{100.0}
\expandafter\def\csname ppt@rev.pol.ni.n.sspro.p1.pass\endcsname{$+$100.0}
\expandafter\def\csname val@rev.pol.ni.n.sspro.p2.lb\endcsname{-0.022}
\expandafter\def\csname cnt@rev.pol.ni.n.sspro.p2.lb\endcsname{0}
\expandafter\def\csname pct@rev.pol.ni.n.sspro.p2.lb\endcsname{$-$2.2}
\expandafter\def\csname ppt@rev.pol.ni.n.sspro.p2.lb\endcsname{$-$2.2}
\expandafter\def\csname val@rev.pol.ni.n.sspro.p2.mean\endcsname{-0.016}
\expandafter\def\csname cnt@rev.pol.ni.n.sspro.p2.mean\endcsname{0}
\expandafter\def\csname pct@rev.pol.ni.n.sspro.p2.mean\endcsname{$-$1.6}
\expandafter\def\csname ppt@rev.pol.ni.n.sspro.p2.mean\endcsname{$-$1.6}
\expandafter\def\csname val@rev.pol.ni.n.sspro.p2.pass\endcsname{0}
\expandafter\def\csname cnt@rev.pol.ni.n.sspro.p2.pass\endcsname{0}
\expandafter\def\csname pct@rev.pol.ni.n.sspro.p2.pass\endcsname{0.0}
\expandafter\def\csname ppt@rev.pol.ni.n.sspro.p2.pass\endcsname{0.0}
\expandafter\def\csname val@rev.pol.ni.n.uiv.p1.lb\endcsname{0.003}
\expandafter\def\csname cnt@rev.pol.ni.n.uiv.p1.lb\endcsname{0}
\expandafter\def\csname pct@rev.pol.ni.n.uiv.p1.lb\endcsname{0.3}
\expandafter\def\csname ppt@rev.pol.ni.n.uiv.p1.lb\endcsname{$+$0.3}
\expandafter\def\csname val@rev.pol.ni.n.uiv.p1.mean\endcsname{0.011}
\expandafter\def\csname cnt@rev.pol.ni.n.uiv.p1.mean\endcsname{0}
\expandafter\def\csname pct@rev.pol.ni.n.uiv.p1.mean\endcsname{1.1}
\expandafter\def\csname ppt@rev.pol.ni.n.uiv.p1.mean\endcsname{$+$1.1}
\expandafter\def\csname val@rev.pol.ni.n.uiv.p1.pass\endcsname{1}
\expandafter\def\csname cnt@rev.pol.ni.n.uiv.p1.pass\endcsname{1}
\expandafter\def\csname pct@rev.pol.ni.n.uiv.p1.pass\endcsname{100.0}
\expandafter\def\csname ppt@rev.pol.ni.n.uiv.p1.pass\endcsname{$+$100.0}
\expandafter\def\csname val@rev.pol.ni.n.uiv.p2.lb\endcsname{-0.008}
\expandafter\def\csname cnt@rev.pol.ni.n.uiv.p2.lb\endcsname{0}
\expandafter\def\csname pct@rev.pol.ni.n.uiv.p2.lb\endcsname{$-$0.8}
\expandafter\def\csname ppt@rev.pol.ni.n.uiv.p2.lb\endcsname{$-$0.8}
\expandafter\def\csname val@rev.pol.ni.n.uiv.p2.mean\endcsname{-0.003}
\expandafter\def\csname cnt@rev.pol.ni.n.uiv.p2.mean\endcsname{0}
\expandafter\def\csname pct@rev.pol.ni.n.uiv.p2.mean\endcsname{$-$0.3}
\expandafter\def\csname ppt@rev.pol.ni.n.uiv.p2.mean\endcsname{$-$0.3}
\expandafter\def\csname val@rev.pol.ni.n.uiv.p2.pass\endcsname{1}
\expandafter\def\csname cnt@rev.pol.ni.n.uiv.p2.pass\endcsname{1}
\expandafter\def\csname pct@rev.pol.ni.n.uiv.p2.pass\endcsname{100.0}
\expandafter\def\csname ppt@rev.pol.ni.n.uiv.p2.pass\endcsname{$+$100.0}
\expandafter\def\csname val@rev.pol.perm.f20.ni.n-pass-signflip\endcsname{3}
\expandafter\def\csname cnt@rev.pol.perm.f20.ni.n-pass-signflip\endcsname{3}
\expandafter\def\csname pct@rev.pol.perm.f20.ni.n-pass-signflip\endcsname{300.0}
\expandafter\def\csname ppt@rev.pol.perm.f20.ni.n-pass-signflip\endcsname{$+$300.0}
\expandafter\def\csname val@rev.pol.perm.f20.ni.n-pass-t\endcsname{3}
\expandafter\def\csname cnt@rev.pol.perm.f20.ni.n-pass-t\endcsname{3}
\expandafter\def\csname pct@rev.pol.perm.f20.ni.n-pass-t\endcsname{300.0}
\expandafter\def\csname ppt@rev.pol.perm.f20.ni.n-pass-t\endcsname{$+$300.0}
\expandafter\def\csname val@rev.pol.perm.f20.signflip-gt.n-rejected\endcsname{3}
\expandafter\def\csname cnt@rev.pol.perm.f20.signflip-gt.n-rejected\endcsname{3}
\expandafter\def\csname pct@rev.pol.perm.f20.signflip-gt.n-rejected\endcsname{300.0}
\expandafter\def\csname ppt@rev.pol.perm.f20.signflip-gt.n-rejected\endcsname{$+$300.0}
\expandafter\def\csname val@rev.pol.perm.f20.sspro.p1.eqcost-lin-vs-best-single.p-signflip\endcsname{0.329}
\expandafter\def\csname cnt@rev.pol.perm.f20.sspro.p1.eqcost-lin-vs-best-single.p-signflip\endcsname{0}
\expandafter\def\csname pct@rev.pol.perm.f20.sspro.p1.eqcost-lin-vs-best-single.p-signflip\endcsname{32.9}
\expandafter\def\csname ppt@rev.pol.perm.f20.sspro.p1.eqcost-lin-vs-best-single.p-signflip\endcsname{$+$32.9}
\expandafter\def\csname val@rev.pol.perm.f20.sspro.p1.eqcost-lin-vs-best-single.p-signflip-lt\endcsname{0.672}
\expandafter\def\csname cnt@rev.pol.perm.f20.sspro.p1.eqcost-lin-vs-best-single.p-signflip-lt\endcsname{1}
\expandafter\def\csname pct@rev.pol.perm.f20.sspro.p1.eqcost-lin-vs-best-single.p-signflip-lt\endcsname{67.2}
\expandafter\def\csname ppt@rev.pol.perm.f20.sspro.p1.eqcost-lin-vs-best-single.p-signflip-lt\endcsname{$+$67.2}
\expandafter\def\csname val@rev.pol.perm.f20.sspro.p1.eqcost-lin-vs-best-single.p-t\endcsname{0.325}
\expandafter\def\csname cnt@rev.pol.perm.f20.sspro.p1.eqcost-lin-vs-best-single.p-t\endcsname{0}
\expandafter\def\csname pct@rev.pol.perm.f20.sspro.p1.eqcost-lin-vs-best-single.p-t\endcsname{32.5}
\expandafter\def\csname ppt@rev.pol.perm.f20.sspro.p1.eqcost-lin-vs-best-single.p-t\endcsname{$+$32.5}
\expandafter\def\csname val@rev.pol.perm.f20.sspro.p1.eqcost-lin-vs-best-single.p-t-lt\endcsname{0.675}
\expandafter\def\csname cnt@rev.pol.perm.f20.sspro.p1.eqcost-lin-vs-best-single.p-t-lt\endcsname{1}
\expandafter\def\csname pct@rev.pol.perm.f20.sspro.p1.eqcost-lin-vs-best-single.p-t-lt\endcsname{67.5}
\expandafter\def\csname ppt@rev.pol.perm.f20.sspro.p1.eqcost-lin-vs-best-single.p-t-lt\endcsname{$+$67.5}
\expandafter\def\csname val@rev.pol.perm.f20.sspro.p1.eqcost-lin-vs-generic-ppi-source.p-signflip\endcsname{0.972}
\expandafter\def\csname cnt@rev.pol.perm.f20.sspro.p1.eqcost-lin-vs-generic-ppi-source.p-signflip\endcsname{1}
\expandafter\def\csname pct@rev.pol.perm.f20.sspro.p1.eqcost-lin-vs-generic-ppi-source.p-signflip\endcsname{97.2}
\expandafter\def\csname ppt@rev.pol.perm.f20.sspro.p1.eqcost-lin-vs-generic-ppi-source.p-signflip\endcsname{$+$97.2}
\expandafter\def\csname val@rev.pol.perm.f20.sspro.p1.eqcost-lin-vs-generic-ppi-source.p-signflip-lt\endcsname{0.063}
\expandafter\def\csname cnt@rev.pol.perm.f20.sspro.p1.eqcost-lin-vs-generic-ppi-source.p-signflip-lt\endcsname{0}
\expandafter\def\csname pct@rev.pol.perm.f20.sspro.p1.eqcost-lin-vs-generic-ppi-source.p-signflip-lt\endcsname{6.3}
\expandafter\def\csname ppt@rev.pol.perm.f20.sspro.p1.eqcost-lin-vs-generic-ppi-source.p-signflip-lt\endcsname{$+$6.3}
\expandafter\def\csname val@rev.pol.perm.f20.sspro.p1.eqcost-lin-vs-generic-ppi-source.p-t\endcsname{0.955}
\expandafter\def\csname cnt@rev.pol.perm.f20.sspro.p1.eqcost-lin-vs-generic-ppi-source.p-t\endcsname{1}
\expandafter\def\csname pct@rev.pol.perm.f20.sspro.p1.eqcost-lin-vs-generic-ppi-source.p-t\endcsname{95.5}
\expandafter\def\csname ppt@rev.pol.perm.f20.sspro.p1.eqcost-lin-vs-generic-ppi-source.p-t\endcsname{$+$95.5}
\expandafter\def\csname val@rev.pol.perm.f20.sspro.p1.eqcost-lin-vs-generic-ppi-source.p-t-lt\endcsname{0.045}
\expandafter\def\csname cnt@rev.pol.perm.f20.sspro.p1.eqcost-lin-vs-generic-ppi-source.p-t-lt\endcsname{0}
\expandafter\def\csname pct@rev.pol.perm.f20.sspro.p1.eqcost-lin-vs-generic-ppi-source.p-t-lt\endcsname{4.5}
\expandafter\def\csname ppt@rev.pol.perm.f20.sspro.p1.eqcost-lin-vs-generic-ppi-source.p-t-lt\endcsname{$+$4.5}
\expandafter\def\csname val@rev.pol.perm.f20.sspro.p1.eqcost-lin-vs-replay-ppi-uniform.p-signflip\endcsname{0.039}
\expandafter\def\csname cnt@rev.pol.perm.f20.sspro.p1.eqcost-lin-vs-replay-ppi-uniform.p-signflip\endcsname{0}
\expandafter\def\csname pct@rev.pol.perm.f20.sspro.p1.eqcost-lin-vs-replay-ppi-uniform.p-signflip\endcsname{3.9}
\expandafter\def\csname ppt@rev.pol.perm.f20.sspro.p1.eqcost-lin-vs-replay-ppi-uniform.p-signflip\endcsname{$+$3.9}
\expandafter\def\csname val@rev.pol.perm.f20.sspro.p1.eqcost-lin-vs-replay-ppi-uniform.p-signflip-lt\endcsname{0.962}
\expandafter\def\csname cnt@rev.pol.perm.f20.sspro.p1.eqcost-lin-vs-replay-ppi-uniform.p-signflip-lt\endcsname{1}
\expandafter\def\csname pct@rev.pol.perm.f20.sspro.p1.eqcost-lin-vs-replay-ppi-uniform.p-signflip-lt\endcsname{96.2}
\expandafter\def\csname ppt@rev.pol.perm.f20.sspro.p1.eqcost-lin-vs-replay-ppi-uniform.p-signflip-lt\endcsname{$+$96.2}
\expandafter\def\csname val@rev.pol.perm.f20.sspro.p1.eqcost-lin-vs-replay-ppi-uniform.p-t\endcsname{0.042}
\expandafter\def\csname cnt@rev.pol.perm.f20.sspro.p1.eqcost-lin-vs-replay-ppi-uniform.p-t\endcsname{0}
\expandafter\def\csname pct@rev.pol.perm.f20.sspro.p1.eqcost-lin-vs-replay-ppi-uniform.p-t\endcsname{4.2}
\expandafter\def\csname ppt@rev.pol.perm.f20.sspro.p1.eqcost-lin-vs-replay-ppi-uniform.p-t\endcsname{$+$4.2}
\expandafter\def\csname val@rev.pol.perm.f20.sspro.p1.eqcost-lin-vs-replay-ppi-uniform.p-t-lt\endcsname{0.958}
\expandafter\def\csname cnt@rev.pol.perm.f20.sspro.p1.eqcost-lin-vs-replay-ppi-uniform.p-t-lt\endcsname{1}
\expandafter\def\csname pct@rev.pol.perm.f20.sspro.p1.eqcost-lin-vs-replay-ppi-uniform.p-t-lt\endcsname{95.8}
\expandafter\def\csname ppt@rev.pol.perm.f20.sspro.p1.eqcost-lin-vs-replay-ppi-uniform.p-t-lt\endcsname{$+$95.8}
\expandafter\def\csname val@rev.pol.perm.f20.sspro.p1.eqcost-lin-vs-uniform-fresh.p-signflip\endcsname{0.972}
\expandafter\def\csname cnt@rev.pol.perm.f20.sspro.p1.eqcost-lin-vs-uniform-fresh.p-signflip\endcsname{1}
\expandafter\def\csname pct@rev.pol.perm.f20.sspro.p1.eqcost-lin-vs-uniform-fresh.p-signflip\endcsname{97.2}
\expandafter\def\csname ppt@rev.pol.perm.f20.sspro.p1.eqcost-lin-vs-uniform-fresh.p-signflip\endcsname{$+$97.2}
\expandafter\def\csname val@rev.pol.perm.f20.sspro.p1.eqcost-lin-vs-uniform-fresh.p-signflip-lt\endcsname{0.063}
\expandafter\def\csname cnt@rev.pol.perm.f20.sspro.p1.eqcost-lin-vs-uniform-fresh.p-signflip-lt\endcsname{0}
\expandafter\def\csname pct@rev.pol.perm.f20.sspro.p1.eqcost-lin-vs-uniform-fresh.p-signflip-lt\endcsname{6.3}
\expandafter\def\csname ppt@rev.pol.perm.f20.sspro.p1.eqcost-lin-vs-uniform-fresh.p-signflip-lt\endcsname{$+$6.3}
\expandafter\def\csname val@rev.pol.perm.f20.sspro.p1.eqcost-lin-vs-uniform-fresh.p-t\endcsname{0.955}
\expandafter\def\csname cnt@rev.pol.perm.f20.sspro.p1.eqcost-lin-vs-uniform-fresh.p-t\endcsname{1}
\expandafter\def\csname pct@rev.pol.perm.f20.sspro.p1.eqcost-lin-vs-uniform-fresh.p-t\endcsname{95.5}
\expandafter\def\csname ppt@rev.pol.perm.f20.sspro.p1.eqcost-lin-vs-uniform-fresh.p-t\endcsname{$+$95.5}
\expandafter\def\csname val@rev.pol.perm.f20.sspro.p1.eqcost-lin-vs-uniform-fresh.p-t-lt\endcsname{0.045}
\expandafter\def\csname cnt@rev.pol.perm.f20.sspro.p1.eqcost-lin-vs-uniform-fresh.p-t-lt\endcsname{0}
\expandafter\def\csname pct@rev.pol.perm.f20.sspro.p1.eqcost-lin-vs-uniform-fresh.p-t-lt\endcsname{4.5}
\expandafter\def\csname ppt@rev.pol.perm.f20.sspro.p1.eqcost-lin-vs-uniform-fresh.p-t-lt\endcsname{$+$4.5}
\expandafter\def\csname val@rev.pol.perm.f20.sspro.p1.rcr-vs-full-recert-ni.p-signflip\endcsname{0.000}
\expandafter\def\csname cnt@rev.pol.perm.f20.sspro.p1.rcr-vs-full-recert-ni.p-signflip\endcsname{0}
\expandafter\def\csname pct@rev.pol.perm.f20.sspro.p1.rcr-vs-full-recert-ni.p-signflip\endcsname{0.0}
\expandafter\def\csname ppt@rev.pol.perm.f20.sspro.p1.rcr-vs-full-recert-ni.p-signflip\endcsname{0.0}
\expandafter\def\csname val@rev.pol.perm.f20.sspro.p1.rcr-vs-full-recert-ni.p-t\endcsname{0.000}
\expandafter\def\csname cnt@rev.pol.perm.f20.sspro.p1.rcr-vs-full-recert-ni.p-t\endcsname{0}
\expandafter\def\csname pct@rev.pol.perm.f20.sspro.p1.rcr-vs-full-recert-ni.p-t\endcsname{0.0}
\expandafter\def\csname ppt@rev.pol.perm.f20.sspro.p1.rcr-vs-full-recert-ni.p-t\endcsname{0.0}
\expandafter\def\csname val@rev.pol.perm.f20.sspro.p1.rcr-vs-generic-ppi-source.p-signflip\endcsname{0.029}
\expandafter\def\csname cnt@rev.pol.perm.f20.sspro.p1.rcr-vs-generic-ppi-source.p-signflip\endcsname{0}
\expandafter\def\csname pct@rev.pol.perm.f20.sspro.p1.rcr-vs-generic-ppi-source.p-signflip\endcsname{2.9}
\expandafter\def\csname ppt@rev.pol.perm.f20.sspro.p1.rcr-vs-generic-ppi-source.p-signflip\endcsname{$+$2.9}
\expandafter\def\csname val@rev.pol.perm.f20.sspro.p1.rcr-vs-generic-ppi-source.p-signflip-holm\endcsname{0.145}
\expandafter\def\csname cnt@rev.pol.perm.f20.sspro.p1.rcr-vs-generic-ppi-source.p-signflip-holm\endcsname{0}
\expandafter\def\csname pct@rev.pol.perm.f20.sspro.p1.rcr-vs-generic-ppi-source.p-signflip-holm\endcsname{14.5}
\expandafter\def\csname ppt@rev.pol.perm.f20.sspro.p1.rcr-vs-generic-ppi-source.p-signflip-holm\endcsname{$+$14.5}
\expandafter\def\csname val@rev.pol.perm.f20.sspro.p1.rcr-vs-generic-ppi-source.p-t\endcsname{0.037}
\expandafter\def\csname cnt@rev.pol.perm.f20.sspro.p1.rcr-vs-generic-ppi-source.p-t\endcsname{0}
\expandafter\def\csname pct@rev.pol.perm.f20.sspro.p1.rcr-vs-generic-ppi-source.p-t\endcsname{3.7}
\expandafter\def\csname ppt@rev.pol.perm.f20.sspro.p1.rcr-vs-generic-ppi-source.p-t\endcsname{$+$3.7}
\expandafter\def\csname val@rev.pol.perm.f20.sspro.p1.rcr-vs-generic-ppi-source.p-t-holm\endcsname{0.134}
\expandafter\def\csname cnt@rev.pol.perm.f20.sspro.p1.rcr-vs-generic-ppi-source.p-t-holm\endcsname{0}
\expandafter\def\csname pct@rev.pol.perm.f20.sspro.p1.rcr-vs-generic-ppi-source.p-t-holm\endcsname{13.4}
\expandafter\def\csname ppt@rev.pol.perm.f20.sspro.p1.rcr-vs-generic-ppi-source.p-t-holm\endcsname{$+$13.4}
\expandafter\def\csname val@rev.pol.perm.f20.sspro.p1.rcr-vs-replay-ppi-uniform.p-signflip\endcsname{0.046}
\expandafter\def\csname cnt@rev.pol.perm.f20.sspro.p1.rcr-vs-replay-ppi-uniform.p-signflip\endcsname{0}
\expandafter\def\csname pct@rev.pol.perm.f20.sspro.p1.rcr-vs-replay-ppi-uniform.p-signflip\endcsname{4.6}
\expandafter\def\csname ppt@rev.pol.perm.f20.sspro.p1.rcr-vs-replay-ppi-uniform.p-signflip\endcsname{$+$4.6}
\expandafter\def\csname val@rev.pol.perm.f20.sspro.p1.rcr-vs-replay-ppi-uniform.p-signflip-holm\endcsname{0.145}
\expandafter\def\csname cnt@rev.pol.perm.f20.sspro.p1.rcr-vs-replay-ppi-uniform.p-signflip-holm\endcsname{0}
\expandafter\def\csname pct@rev.pol.perm.f20.sspro.p1.rcr-vs-replay-ppi-uniform.p-signflip-holm\endcsname{14.5}
\expandafter\def\csname ppt@rev.pol.perm.f20.sspro.p1.rcr-vs-replay-ppi-uniform.p-signflip-holm\endcsname{$+$14.5}
\expandafter\def\csname val@rev.pol.perm.f20.sspro.p1.rcr-vs-replay-ppi-uniform.p-t\endcsname{0.042}
\expandafter\def\csname cnt@rev.pol.perm.f20.sspro.p1.rcr-vs-replay-ppi-uniform.p-t\endcsname{0}
\expandafter\def\csname pct@rev.pol.perm.f20.sspro.p1.rcr-vs-replay-ppi-uniform.p-t\endcsname{4.2}
\expandafter\def\csname ppt@rev.pol.perm.f20.sspro.p1.rcr-vs-replay-ppi-uniform.p-t\endcsname{$+$4.2}
\expandafter\def\csname val@rev.pol.perm.f20.sspro.p1.rcr-vs-replay-ppi-uniform.p-t-holm\endcsname{0.134}
\expandafter\def\csname cnt@rev.pol.perm.f20.sspro.p1.rcr-vs-replay-ppi-uniform.p-t-holm\endcsname{0}
\expandafter\def\csname pct@rev.pol.perm.f20.sspro.p1.rcr-vs-replay-ppi-uniform.p-t-holm\endcsname{13.4}
\expandafter\def\csname ppt@rev.pol.perm.f20.sspro.p1.rcr-vs-replay-ppi-uniform.p-t-holm\endcsname{$+$13.4}
\expandafter\def\csname val@rev.pol.perm.f20.sspro.p2.eqcost-lin-vs-best-single.p-signflip\endcsname{0.962}
\expandafter\def\csname cnt@rev.pol.perm.f20.sspro.p2.eqcost-lin-vs-best-single.p-signflip\endcsname{1}
\expandafter\def\csname pct@rev.pol.perm.f20.sspro.p2.eqcost-lin-vs-best-single.p-signflip\endcsname{96.2}
\expandafter\def\csname ppt@rev.pol.perm.f20.sspro.p2.eqcost-lin-vs-best-single.p-signflip\endcsname{$+$96.2}
\expandafter\def\csname val@rev.pol.perm.f20.sspro.p2.eqcost-lin-vs-best-single.p-signflip-lt\endcsname{0.039}
\expandafter\def\csname cnt@rev.pol.perm.f20.sspro.p2.eqcost-lin-vs-best-single.p-signflip-lt\endcsname{0}
\expandafter\def\csname pct@rev.pol.perm.f20.sspro.p2.eqcost-lin-vs-best-single.p-signflip-lt\endcsname{3.9}
\expandafter\def\csname ppt@rev.pol.perm.f20.sspro.p2.eqcost-lin-vs-best-single.p-signflip-lt\endcsname{$+$3.9}
\expandafter\def\csname val@rev.pol.perm.f20.sspro.p2.eqcost-lin-vs-best-single.p-t\endcsname{0.962}
\expandafter\def\csname cnt@rev.pol.perm.f20.sspro.p2.eqcost-lin-vs-best-single.p-t\endcsname{1}
\expandafter\def\csname pct@rev.pol.perm.f20.sspro.p2.eqcost-lin-vs-best-single.p-t\endcsname{96.2}
\expandafter\def\csname ppt@rev.pol.perm.f20.sspro.p2.eqcost-lin-vs-best-single.p-t\endcsname{$+$96.2}
\expandafter\def\csname val@rev.pol.perm.f20.sspro.p2.eqcost-lin-vs-best-single.p-t-lt\endcsname{0.038}
\expandafter\def\csname cnt@rev.pol.perm.f20.sspro.p2.eqcost-lin-vs-best-single.p-t-lt\endcsname{0}
\expandafter\def\csname pct@rev.pol.perm.f20.sspro.p2.eqcost-lin-vs-best-single.p-t-lt\endcsname{3.8}
\expandafter\def\csname ppt@rev.pol.perm.f20.sspro.p2.eqcost-lin-vs-best-single.p-t-lt\endcsname{$+$3.8}
\expandafter\def\csname val@rev.pol.perm.f20.sspro.p2.eqcost-lin-vs-generic-ppi-source.p-signflip\endcsname{0.999}
\expandafter\def\csname cnt@rev.pol.perm.f20.sspro.p2.eqcost-lin-vs-generic-ppi-source.p-signflip\endcsname{1}
\expandafter\def\csname pct@rev.pol.perm.f20.sspro.p2.eqcost-lin-vs-generic-ppi-source.p-signflip\endcsname{99.9}
\expandafter\def\csname ppt@rev.pol.perm.f20.sspro.p2.eqcost-lin-vs-generic-ppi-source.p-signflip\endcsname{$+$99.9}
\expandafter\def\csname val@rev.pol.perm.f20.sspro.p2.eqcost-lin-vs-generic-ppi-source.p-signflip-lt\endcsname{0.002}
\expandafter\def\csname cnt@rev.pol.perm.f20.sspro.p2.eqcost-lin-vs-generic-ppi-source.p-signflip-lt\endcsname{0}
\expandafter\def\csname pct@rev.pol.perm.f20.sspro.p2.eqcost-lin-vs-generic-ppi-source.p-signflip-lt\endcsname{0.2}
\expandafter\def\csname ppt@rev.pol.perm.f20.sspro.p2.eqcost-lin-vs-generic-ppi-source.p-signflip-lt\endcsname{$+$0.2}
\expandafter\def\csname val@rev.pol.perm.f20.sspro.p2.eqcost-lin-vs-generic-ppi-source.p-t\endcsname{0.998}
\expandafter\def\csname cnt@rev.pol.perm.f20.sspro.p2.eqcost-lin-vs-generic-ppi-source.p-t\endcsname{1}
\expandafter\def\csname pct@rev.pol.perm.f20.sspro.p2.eqcost-lin-vs-generic-ppi-source.p-t\endcsname{99.8}
\expandafter\def\csname ppt@rev.pol.perm.f20.sspro.p2.eqcost-lin-vs-generic-ppi-source.p-t\endcsname{$+$99.8}
\expandafter\def\csname val@rev.pol.perm.f20.sspro.p2.eqcost-lin-vs-generic-ppi-source.p-t-lt\endcsname{0.002}
\expandafter\def\csname cnt@rev.pol.perm.f20.sspro.p2.eqcost-lin-vs-generic-ppi-source.p-t-lt\endcsname{0}
\expandafter\def\csname pct@rev.pol.perm.f20.sspro.p2.eqcost-lin-vs-generic-ppi-source.p-t-lt\endcsname{0.2}
\expandafter\def\csname ppt@rev.pol.perm.f20.sspro.p2.eqcost-lin-vs-generic-ppi-source.p-t-lt\endcsname{$+$0.2}
\expandafter\def\csname val@rev.pol.perm.f20.sspro.p2.eqcost-lin-vs-replay-ppi-uniform.p-signflip\endcsname{0.001}
\expandafter\def\csname cnt@rev.pol.perm.f20.sspro.p2.eqcost-lin-vs-replay-ppi-uniform.p-signflip\endcsname{0}
\expandafter\def\csname pct@rev.pol.perm.f20.sspro.p2.eqcost-lin-vs-replay-ppi-uniform.p-signflip\endcsname{0.1}
\expandafter\def\csname ppt@rev.pol.perm.f20.sspro.p2.eqcost-lin-vs-replay-ppi-uniform.p-signflip\endcsname{$+$0.1}
\expandafter\def\csname val@rev.pol.perm.f20.sspro.p2.eqcost-lin-vs-replay-ppi-uniform.p-signflip-lt\endcsname{1.000}
\expandafter\def\csname cnt@rev.pol.perm.f20.sspro.p2.eqcost-lin-vs-replay-ppi-uniform.p-signflip-lt\endcsname{1}
\expandafter\def\csname pct@rev.pol.perm.f20.sspro.p2.eqcost-lin-vs-replay-ppi-uniform.p-signflip-lt\endcsname{100.0}
\expandafter\def\csname ppt@rev.pol.perm.f20.sspro.p2.eqcost-lin-vs-replay-ppi-uniform.p-signflip-lt\endcsname{$+$100.0}
\expandafter\def\csname val@rev.pol.perm.f20.sspro.p2.eqcost-lin-vs-replay-ppi-uniform.p-t\endcsname{0.001}
\expandafter\def\csname cnt@rev.pol.perm.f20.sspro.p2.eqcost-lin-vs-replay-ppi-uniform.p-t\endcsname{0}
\expandafter\def\csname pct@rev.pol.perm.f20.sspro.p2.eqcost-lin-vs-replay-ppi-uniform.p-t\endcsname{0.1}
\expandafter\def\csname ppt@rev.pol.perm.f20.sspro.p2.eqcost-lin-vs-replay-ppi-uniform.p-t\endcsname{$+$0.1}
\expandafter\def\csname val@rev.pol.perm.f20.sspro.p2.eqcost-lin-vs-replay-ppi-uniform.p-t-lt\endcsname{0.999}
\expandafter\def\csname cnt@rev.pol.perm.f20.sspro.p2.eqcost-lin-vs-replay-ppi-uniform.p-t-lt\endcsname{1}
\expandafter\def\csname pct@rev.pol.perm.f20.sspro.p2.eqcost-lin-vs-replay-ppi-uniform.p-t-lt\endcsname{99.9}
\expandafter\def\csname ppt@rev.pol.perm.f20.sspro.p2.eqcost-lin-vs-replay-ppi-uniform.p-t-lt\endcsname{$+$99.9}
\expandafter\def\csname val@rev.pol.perm.f20.sspro.p2.eqcost-lin-vs-uniform-fresh.p-signflip\endcsname{1.000}
\expandafter\def\csname cnt@rev.pol.perm.f20.sspro.p2.eqcost-lin-vs-uniform-fresh.p-signflip\endcsname{1}
\expandafter\def\csname pct@rev.pol.perm.f20.sspro.p2.eqcost-lin-vs-uniform-fresh.p-signflip\endcsname{100.0}
\expandafter\def\csname ppt@rev.pol.perm.f20.sspro.p2.eqcost-lin-vs-uniform-fresh.p-signflip\endcsname{$+$100.0}
\expandafter\def\csname val@rev.pol.perm.f20.sspro.p2.eqcost-lin-vs-uniform-fresh.p-signflip-lt\endcsname{0.001}
\expandafter\def\csname cnt@rev.pol.perm.f20.sspro.p2.eqcost-lin-vs-uniform-fresh.p-signflip-lt\endcsname{0}
\expandafter\def\csname pct@rev.pol.perm.f20.sspro.p2.eqcost-lin-vs-uniform-fresh.p-signflip-lt\endcsname{0.1}
\expandafter\def\csname ppt@rev.pol.perm.f20.sspro.p2.eqcost-lin-vs-uniform-fresh.p-signflip-lt\endcsname{$+$0.1}
\expandafter\def\csname val@rev.pol.perm.f20.sspro.p2.eqcost-lin-vs-uniform-fresh.p-t\endcsname{0.999}
\expandafter\def\csname cnt@rev.pol.perm.f20.sspro.p2.eqcost-lin-vs-uniform-fresh.p-t\endcsname{1}
\expandafter\def\csname pct@rev.pol.perm.f20.sspro.p2.eqcost-lin-vs-uniform-fresh.p-t\endcsname{99.9}
\expandafter\def\csname ppt@rev.pol.perm.f20.sspro.p2.eqcost-lin-vs-uniform-fresh.p-t\endcsname{$+$99.9}
\expandafter\def\csname val@rev.pol.perm.f20.sspro.p2.eqcost-lin-vs-uniform-fresh.p-t-lt\endcsname{0.001}
\expandafter\def\csname cnt@rev.pol.perm.f20.sspro.p2.eqcost-lin-vs-uniform-fresh.p-t-lt\endcsname{0}
\expandafter\def\csname pct@rev.pol.perm.f20.sspro.p2.eqcost-lin-vs-uniform-fresh.p-t-lt\endcsname{0.1}
\expandafter\def\csname ppt@rev.pol.perm.f20.sspro.p2.eqcost-lin-vs-uniform-fresh.p-t-lt\endcsname{$+$0.1}
\expandafter\def\csname val@rev.pol.perm.f20.sspro.p2.rcr-vs-full-recert-ni.p-signflip\endcsname{0.693}
\expandafter\def\csname cnt@rev.pol.perm.f20.sspro.p2.rcr-vs-full-recert-ni.p-signflip\endcsname{1}
\expandafter\def\csname pct@rev.pol.perm.f20.sspro.p2.rcr-vs-full-recert-ni.p-signflip\endcsname{69.3}
\expandafter\def\csname ppt@rev.pol.perm.f20.sspro.p2.rcr-vs-full-recert-ni.p-signflip\endcsname{$+$69.3}
\expandafter\def\csname val@rev.pol.perm.f20.sspro.p2.rcr-vs-full-recert-ni.p-t\endcsname{0.691}
\expandafter\def\csname cnt@rev.pol.perm.f20.sspro.p2.rcr-vs-full-recert-ni.p-t\endcsname{1}
\expandafter\def\csname pct@rev.pol.perm.f20.sspro.p2.rcr-vs-full-recert-ni.p-t\endcsname{69.1}
\expandafter\def\csname ppt@rev.pol.perm.f20.sspro.p2.rcr-vs-full-recert-ni.p-t\endcsname{$+$69.1}
\expandafter\def\csname val@rev.pol.perm.f20.sspro.p2.rcr-vs-generic-ppi-source.p-signflip\endcsname{0.001}
\expandafter\def\csname cnt@rev.pol.perm.f20.sspro.p2.rcr-vs-generic-ppi-source.p-signflip\endcsname{0}
\expandafter\def\csname pct@rev.pol.perm.f20.sspro.p2.rcr-vs-generic-ppi-source.p-signflip\endcsname{0.1}
\expandafter\def\csname ppt@rev.pol.perm.f20.sspro.p2.rcr-vs-generic-ppi-source.p-signflip\endcsname{$+$0.1}
\expandafter\def\csname val@rev.pol.perm.f20.sspro.p2.rcr-vs-generic-ppi-source.p-signflip-holm\endcsname{0.007}
\expandafter\def\csname cnt@rev.pol.perm.f20.sspro.p2.rcr-vs-generic-ppi-source.p-signflip-holm\endcsname{0}
\expandafter\def\csname pct@rev.pol.perm.f20.sspro.p2.rcr-vs-generic-ppi-source.p-signflip-holm\endcsname{0.7}
\expandafter\def\csname ppt@rev.pol.perm.f20.sspro.p2.rcr-vs-generic-ppi-source.p-signflip-holm\endcsname{$+$0.7}
\expandafter\def\csname val@rev.pol.perm.f20.sspro.p2.rcr-vs-generic-ppi-source.p-t\endcsname{0.001}
\expandafter\def\csname cnt@rev.pol.perm.f20.sspro.p2.rcr-vs-generic-ppi-source.p-t\endcsname{0}
\expandafter\def\csname pct@rev.pol.perm.f20.sspro.p2.rcr-vs-generic-ppi-source.p-t\endcsname{0.1}
\expandafter\def\csname ppt@rev.pol.perm.f20.sspro.p2.rcr-vs-generic-ppi-source.p-t\endcsname{$+$0.1}
\expandafter\def\csname val@rev.pol.perm.f20.sspro.p2.rcr-vs-generic-ppi-source.p-t-holm\endcsname{0.006}
\expandafter\def\csname cnt@rev.pol.perm.f20.sspro.p2.rcr-vs-generic-ppi-source.p-t-holm\endcsname{0}
\expandafter\def\csname pct@rev.pol.perm.f20.sspro.p2.rcr-vs-generic-ppi-source.p-t-holm\endcsname{0.6}
\expandafter\def\csname ppt@rev.pol.perm.f20.sspro.p2.rcr-vs-generic-ppi-source.p-t-holm\endcsname{$+$0.6}
\expandafter\def\csname val@rev.pol.perm.f20.sspro.p2.rcr-vs-replay-ppi-uniform.p-signflip\endcsname{0.001}
\expandafter\def\csname cnt@rev.pol.perm.f20.sspro.p2.rcr-vs-replay-ppi-uniform.p-signflip\endcsname{0}
\expandafter\def\csname pct@rev.pol.perm.f20.sspro.p2.rcr-vs-replay-ppi-uniform.p-signflip\endcsname{0.1}
\expandafter\def\csname ppt@rev.pol.perm.f20.sspro.p2.rcr-vs-replay-ppi-uniform.p-signflip\endcsname{$+$0.1}
\expandafter\def\csname val@rev.pol.perm.f20.sspro.p2.rcr-vs-replay-ppi-uniform.p-signflip-holm\endcsname{0.007}
\expandafter\def\csname cnt@rev.pol.perm.f20.sspro.p2.rcr-vs-replay-ppi-uniform.p-signflip-holm\endcsname{0}
\expandafter\def\csname pct@rev.pol.perm.f20.sspro.p2.rcr-vs-replay-ppi-uniform.p-signflip-holm\endcsname{0.7}
\expandafter\def\csname ppt@rev.pol.perm.f20.sspro.p2.rcr-vs-replay-ppi-uniform.p-signflip-holm\endcsname{$+$0.7}
\expandafter\def\csname val@rev.pol.perm.f20.sspro.p2.rcr-vs-replay-ppi-uniform.p-t\endcsname{0.001}
\expandafter\def\csname cnt@rev.pol.perm.f20.sspro.p2.rcr-vs-replay-ppi-uniform.p-t\endcsname{0}
\expandafter\def\csname pct@rev.pol.perm.f20.sspro.p2.rcr-vs-replay-ppi-uniform.p-t\endcsname{0.1}
\expandafter\def\csname ppt@rev.pol.perm.f20.sspro.p2.rcr-vs-replay-ppi-uniform.p-t\endcsname{$+$0.1}
\expandafter\def\csname val@rev.pol.perm.f20.sspro.p2.rcr-vs-replay-ppi-uniform.p-t-holm\endcsname{0.006}
\expandafter\def\csname cnt@rev.pol.perm.f20.sspro.p2.rcr-vs-replay-ppi-uniform.p-t-holm\endcsname{0}
\expandafter\def\csname pct@rev.pol.perm.f20.sspro.p2.rcr-vs-replay-ppi-uniform.p-t-holm\endcsname{0.6}
\expandafter\def\csname ppt@rev.pol.perm.f20.sspro.p2.rcr-vs-replay-ppi-uniform.p-t-holm\endcsname{$+$0.6}
\expandafter\def\csname val@rev.pol.perm.f20.ttest-gt.n-rejected\endcsname{3}
\expandafter\def\csname cnt@rev.pol.perm.f20.ttest-gt.n-rejected\endcsname{3}
\expandafter\def\csname pct@rev.pol.perm.f20.ttest-gt.n-rejected\endcsname{300.0}
\expandafter\def\csname ppt@rev.pol.perm.f20.ttest-gt.n-rejected\endcsname{$+$300.0}
\expandafter\def\csname val@rev.pol.perm.f20.uiv.p1.eqcost-lin-vs-best-single.p-signflip\endcsname{1.000}
\expandafter\def\csname cnt@rev.pol.perm.f20.uiv.p1.eqcost-lin-vs-best-single.p-signflip\endcsname{1}
\expandafter\def\csname pct@rev.pol.perm.f20.uiv.p1.eqcost-lin-vs-best-single.p-signflip\endcsname{100.0}
\expandafter\def\csname ppt@rev.pol.perm.f20.uiv.p1.eqcost-lin-vs-best-single.p-signflip\endcsname{$+$100.0}
\expandafter\def\csname val@rev.pol.perm.f20.uiv.p1.eqcost-lin-vs-best-single.p-signflip-lt\endcsname{0.000}
\expandafter\def\csname cnt@rev.pol.perm.f20.uiv.p1.eqcost-lin-vs-best-single.p-signflip-lt\endcsname{0}
\expandafter\def\csname pct@rev.pol.perm.f20.uiv.p1.eqcost-lin-vs-best-single.p-signflip-lt\endcsname{0.0}
\expandafter\def\csname ppt@rev.pol.perm.f20.uiv.p1.eqcost-lin-vs-best-single.p-signflip-lt\endcsname{0.0}
\expandafter\def\csname val@rev.pol.perm.f20.uiv.p1.eqcost-lin-vs-best-single.p-t\endcsname{1.000}
\expandafter\def\csname cnt@rev.pol.perm.f20.uiv.p1.eqcost-lin-vs-best-single.p-t\endcsname{1}
\expandafter\def\csname pct@rev.pol.perm.f20.uiv.p1.eqcost-lin-vs-best-single.p-t\endcsname{100.0}
\expandafter\def\csname ppt@rev.pol.perm.f20.uiv.p1.eqcost-lin-vs-best-single.p-t\endcsname{$+$100.0}
\expandafter\def\csname val@rev.pol.perm.f20.uiv.p1.eqcost-lin-vs-best-single.p-t-lt\endcsname{0.000}
\expandafter\def\csname cnt@rev.pol.perm.f20.uiv.p1.eqcost-lin-vs-best-single.p-t-lt\endcsname{0}
\expandafter\def\csname pct@rev.pol.perm.f20.uiv.p1.eqcost-lin-vs-best-single.p-t-lt\endcsname{0.0}
\expandafter\def\csname ppt@rev.pol.perm.f20.uiv.p1.eqcost-lin-vs-best-single.p-t-lt\endcsname{0.0}
\expandafter\def\csname val@rev.pol.perm.f20.uiv.p1.eqcost-lin-vs-generic-ppi-source.p-signflip\endcsname{0.000}
\expandafter\def\csname cnt@rev.pol.perm.f20.uiv.p1.eqcost-lin-vs-generic-ppi-source.p-signflip\endcsname{0}
\expandafter\def\csname pct@rev.pol.perm.f20.uiv.p1.eqcost-lin-vs-generic-ppi-source.p-signflip\endcsname{0.0}
\expandafter\def\csname ppt@rev.pol.perm.f20.uiv.p1.eqcost-lin-vs-generic-ppi-source.p-signflip\endcsname{0.0}
\expandafter\def\csname val@rev.pol.perm.f20.uiv.p1.eqcost-lin-vs-generic-ppi-source.p-signflip-lt\endcsname{1.000}
\expandafter\def\csname cnt@rev.pol.perm.f20.uiv.p1.eqcost-lin-vs-generic-ppi-source.p-signflip-lt\endcsname{1}
\expandafter\def\csname pct@rev.pol.perm.f20.uiv.p1.eqcost-lin-vs-generic-ppi-source.p-signflip-lt\endcsname{100.0}
\expandafter\def\csname ppt@rev.pol.perm.f20.uiv.p1.eqcost-lin-vs-generic-ppi-source.p-signflip-lt\endcsname{$+$100.0}
\expandafter\def\csname val@rev.pol.perm.f20.uiv.p1.eqcost-lin-vs-generic-ppi-source.p-t\endcsname{0.000}
\expandafter\def\csname cnt@rev.pol.perm.f20.uiv.p1.eqcost-lin-vs-generic-ppi-source.p-t\endcsname{0}
\expandafter\def\csname pct@rev.pol.perm.f20.uiv.p1.eqcost-lin-vs-generic-ppi-source.p-t\endcsname{0.0}
\expandafter\def\csname ppt@rev.pol.perm.f20.uiv.p1.eqcost-lin-vs-generic-ppi-source.p-t\endcsname{0.0}
\expandafter\def\csname val@rev.pol.perm.f20.uiv.p1.eqcost-lin-vs-generic-ppi-source.p-t-lt\endcsname{1.000}
\expandafter\def\csname cnt@rev.pol.perm.f20.uiv.p1.eqcost-lin-vs-generic-ppi-source.p-t-lt\endcsname{1}
\expandafter\def\csname pct@rev.pol.perm.f20.uiv.p1.eqcost-lin-vs-generic-ppi-source.p-t-lt\endcsname{100.0}
\expandafter\def\csname ppt@rev.pol.perm.f20.uiv.p1.eqcost-lin-vs-generic-ppi-source.p-t-lt\endcsname{$+$100.0}
\expandafter\def\csname val@rev.pol.perm.f20.uiv.p1.eqcost-lin-vs-replay-ppi-uniform.p-signflip\endcsname{0.221}
\expandafter\def\csname cnt@rev.pol.perm.f20.uiv.p1.eqcost-lin-vs-replay-ppi-uniform.p-signflip\endcsname{0}
\expandafter\def\csname pct@rev.pol.perm.f20.uiv.p1.eqcost-lin-vs-replay-ppi-uniform.p-signflip\endcsname{22.1}
\expandafter\def\csname ppt@rev.pol.perm.f20.uiv.p1.eqcost-lin-vs-replay-ppi-uniform.p-signflip\endcsname{$+$22.1}
\expandafter\def\csname val@rev.pol.perm.f20.uiv.p1.eqcost-lin-vs-replay-ppi-uniform.p-signflip-lt\endcsname{0.785}
\expandafter\def\csname cnt@rev.pol.perm.f20.uiv.p1.eqcost-lin-vs-replay-ppi-uniform.p-signflip-lt\endcsname{1}
\expandafter\def\csname pct@rev.pol.perm.f20.uiv.p1.eqcost-lin-vs-replay-ppi-uniform.p-signflip-lt\endcsname{78.5}
\expandafter\def\csname ppt@rev.pol.perm.f20.uiv.p1.eqcost-lin-vs-replay-ppi-uniform.p-signflip-lt\endcsname{$+$78.5}
\expandafter\def\csname val@rev.pol.perm.f20.uiv.p1.eqcost-lin-vs-replay-ppi-uniform.p-t\endcsname{0.182}
\expandafter\def\csname cnt@rev.pol.perm.f20.uiv.p1.eqcost-lin-vs-replay-ppi-uniform.p-t\endcsname{0}
\expandafter\def\csname pct@rev.pol.perm.f20.uiv.p1.eqcost-lin-vs-replay-ppi-uniform.p-t\endcsname{18.2}
\expandafter\def\csname ppt@rev.pol.perm.f20.uiv.p1.eqcost-lin-vs-replay-ppi-uniform.p-t\endcsname{$+$18.2}
\expandafter\def\csname val@rev.pol.perm.f20.uiv.p1.eqcost-lin-vs-replay-ppi-uniform.p-t-lt\endcsname{0.818}
\expandafter\def\csname cnt@rev.pol.perm.f20.uiv.p1.eqcost-lin-vs-replay-ppi-uniform.p-t-lt\endcsname{1}
\expandafter\def\csname pct@rev.pol.perm.f20.uiv.p1.eqcost-lin-vs-replay-ppi-uniform.p-t-lt\endcsname{81.8}
\expandafter\def\csname ppt@rev.pol.perm.f20.uiv.p1.eqcost-lin-vs-replay-ppi-uniform.p-t-lt\endcsname{$+$81.8}
\expandafter\def\csname val@rev.pol.perm.f20.uiv.p1.eqcost-lin-vs-uniform-fresh.p-signflip\endcsname{0.222}
\expandafter\def\csname cnt@rev.pol.perm.f20.uiv.p1.eqcost-lin-vs-uniform-fresh.p-signflip\endcsname{0}
\expandafter\def\csname pct@rev.pol.perm.f20.uiv.p1.eqcost-lin-vs-uniform-fresh.p-signflip\endcsname{22.2}
\expandafter\def\csname ppt@rev.pol.perm.f20.uiv.p1.eqcost-lin-vs-uniform-fresh.p-signflip\endcsname{$+$22.2}
\expandafter\def\csname val@rev.pol.perm.f20.uiv.p1.eqcost-lin-vs-uniform-fresh.p-signflip-lt\endcsname{0.778}
\expandafter\def\csname cnt@rev.pol.perm.f20.uiv.p1.eqcost-lin-vs-uniform-fresh.p-signflip-lt\endcsname{1}
\expandafter\def\csname pct@rev.pol.perm.f20.uiv.p1.eqcost-lin-vs-uniform-fresh.p-signflip-lt\endcsname{77.8}
\expandafter\def\csname ppt@rev.pol.perm.f20.uiv.p1.eqcost-lin-vs-uniform-fresh.p-signflip-lt\endcsname{$+$77.8}
\expandafter\def\csname val@rev.pol.perm.f20.uiv.p1.eqcost-lin-vs-uniform-fresh.p-t\endcsname{0.221}
\expandafter\def\csname cnt@rev.pol.perm.f20.uiv.p1.eqcost-lin-vs-uniform-fresh.p-t\endcsname{0}
\expandafter\def\csname pct@rev.pol.perm.f20.uiv.p1.eqcost-lin-vs-uniform-fresh.p-t\endcsname{22.1}
\expandafter\def\csname ppt@rev.pol.perm.f20.uiv.p1.eqcost-lin-vs-uniform-fresh.p-t\endcsname{$+$22.1}
\expandafter\def\csname val@rev.pol.perm.f20.uiv.p1.eqcost-lin-vs-uniform-fresh.p-t-lt\endcsname{0.779}
\expandafter\def\csname cnt@rev.pol.perm.f20.uiv.p1.eqcost-lin-vs-uniform-fresh.p-t-lt\endcsname{1}
\expandafter\def\csname pct@rev.pol.perm.f20.uiv.p1.eqcost-lin-vs-uniform-fresh.p-t-lt\endcsname{77.9}
\expandafter\def\csname ppt@rev.pol.perm.f20.uiv.p1.eqcost-lin-vs-uniform-fresh.p-t-lt\endcsname{$+$77.9}
\expandafter\def\csname val@rev.pol.perm.f20.uiv.p1.rcr-vs-full-recert-ni.p-signflip\endcsname{0.000}
\expandafter\def\csname cnt@rev.pol.perm.f20.uiv.p1.rcr-vs-full-recert-ni.p-signflip\endcsname{0}
\expandafter\def\csname pct@rev.pol.perm.f20.uiv.p1.rcr-vs-full-recert-ni.p-signflip\endcsname{0.0}
\expandafter\def\csname ppt@rev.pol.perm.f20.uiv.p1.rcr-vs-full-recert-ni.p-signflip\endcsname{0.0}
\expandafter\def\csname val@rev.pol.perm.f20.uiv.p1.rcr-vs-full-recert-ni.p-t\endcsname{0.000}
\expandafter\def\csname cnt@rev.pol.perm.f20.uiv.p1.rcr-vs-full-recert-ni.p-t\endcsname{0}
\expandafter\def\csname pct@rev.pol.perm.f20.uiv.p1.rcr-vs-full-recert-ni.p-t\endcsname{0.0}
\expandafter\def\csname ppt@rev.pol.perm.f20.uiv.p1.rcr-vs-full-recert-ni.p-t\endcsname{0.0}
\expandafter\def\csname val@rev.pol.perm.f20.uiv.p1.rcr-vs-generic-ppi-source.p-signflip\endcsname{0.000}
\expandafter\def\csname cnt@rev.pol.perm.f20.uiv.p1.rcr-vs-generic-ppi-source.p-signflip\endcsname{0}
\expandafter\def\csname pct@rev.pol.perm.f20.uiv.p1.rcr-vs-generic-ppi-source.p-signflip\endcsname{0.0}
\expandafter\def\csname ppt@rev.pol.perm.f20.uiv.p1.rcr-vs-generic-ppi-source.p-signflip\endcsname{0.0}
\expandafter\def\csname val@rev.pol.perm.f20.uiv.p1.rcr-vs-generic-ppi-source.p-signflip-holm\endcsname{0.002}
\expandafter\def\csname cnt@rev.pol.perm.f20.uiv.p1.rcr-vs-generic-ppi-source.p-signflip-holm\endcsname{0}
\expandafter\def\csname pct@rev.pol.perm.f20.uiv.p1.rcr-vs-generic-ppi-source.p-signflip-holm\endcsname{0.2}
\expandafter\def\csname ppt@rev.pol.perm.f20.uiv.p1.rcr-vs-generic-ppi-source.p-signflip-holm\endcsname{$+$0.2}
\expandafter\def\csname val@rev.pol.perm.f20.uiv.p1.rcr-vs-generic-ppi-source.p-t\endcsname{0.000}
\expandafter\def\csname cnt@rev.pol.perm.f20.uiv.p1.rcr-vs-generic-ppi-source.p-t\endcsname{0}
\expandafter\def\csname pct@rev.pol.perm.f20.uiv.p1.rcr-vs-generic-ppi-source.p-t\endcsname{0.0}
\expandafter\def\csname ppt@rev.pol.perm.f20.uiv.p1.rcr-vs-generic-ppi-source.p-t\endcsname{0.0}
\expandafter\def\csname val@rev.pol.perm.f20.uiv.p1.rcr-vs-generic-ppi-source.p-t-holm\endcsname{0.004}
\expandafter\def\csname cnt@rev.pol.perm.f20.uiv.p1.rcr-vs-generic-ppi-source.p-t-holm\endcsname{0}
\expandafter\def\csname pct@rev.pol.perm.f20.uiv.p1.rcr-vs-generic-ppi-source.p-t-holm\endcsname{0.4}
\expandafter\def\csname ppt@rev.pol.perm.f20.uiv.p1.rcr-vs-generic-ppi-source.p-t-holm\endcsname{$+$0.4}
\expandafter\def\csname val@rev.pol.perm.f20.uiv.p1.rcr-vs-replay-ppi-uniform.p-signflip\endcsname{0.232}
\expandafter\def\csname cnt@rev.pol.perm.f20.uiv.p1.rcr-vs-replay-ppi-uniform.p-signflip\endcsname{0}
\expandafter\def\csname pct@rev.pol.perm.f20.uiv.p1.rcr-vs-replay-ppi-uniform.p-signflip\endcsname{23.2}
\expandafter\def\csname ppt@rev.pol.perm.f20.uiv.p1.rcr-vs-replay-ppi-uniform.p-signflip\endcsname{$+$23.2}
\expandafter\def\csname val@rev.pol.perm.f20.uiv.p1.rcr-vs-replay-ppi-uniform.p-signflip-holm\endcsname{0.232}
\expandafter\def\csname cnt@rev.pol.perm.f20.uiv.p1.rcr-vs-replay-ppi-uniform.p-signflip-holm\endcsname{0}
\expandafter\def\csname pct@rev.pol.perm.f20.uiv.p1.rcr-vs-replay-ppi-uniform.p-signflip-holm\endcsname{23.2}
\expandafter\def\csname ppt@rev.pol.perm.f20.uiv.p1.rcr-vs-replay-ppi-uniform.p-signflip-holm\endcsname{$+$23.2}
\expandafter\def\csname val@rev.pol.perm.f20.uiv.p1.rcr-vs-replay-ppi-uniform.p-t\endcsname{0.182}
\expandafter\def\csname cnt@rev.pol.perm.f20.uiv.p1.rcr-vs-replay-ppi-uniform.p-t\endcsname{0}
\expandafter\def\csname pct@rev.pol.perm.f20.uiv.p1.rcr-vs-replay-ppi-uniform.p-t\endcsname{18.2}
\expandafter\def\csname ppt@rev.pol.perm.f20.uiv.p1.rcr-vs-replay-ppi-uniform.p-t\endcsname{$+$18.2}
\expandafter\def\csname val@rev.pol.perm.f20.uiv.p1.rcr-vs-replay-ppi-uniform.p-t-holm\endcsname{0.182}
\expandafter\def\csname cnt@rev.pol.perm.f20.uiv.p1.rcr-vs-replay-ppi-uniform.p-t-holm\endcsname{0}
\expandafter\def\csname pct@rev.pol.perm.f20.uiv.p1.rcr-vs-replay-ppi-uniform.p-t-holm\endcsname{18.2}
\expandafter\def\csname ppt@rev.pol.perm.f20.uiv.p1.rcr-vs-replay-ppi-uniform.p-t-holm\endcsname{$+$18.2}
\expandafter\def\csname val@rev.pol.perm.f20.uiv.p2.eqcost-lin-vs-best-single.p-signflip\endcsname{0.252}
\expandafter\def\csname cnt@rev.pol.perm.f20.uiv.p2.eqcost-lin-vs-best-single.p-signflip\endcsname{0}
\expandafter\def\csname pct@rev.pol.perm.f20.uiv.p2.eqcost-lin-vs-best-single.p-signflip\endcsname{25.2}
\expandafter\def\csname ppt@rev.pol.perm.f20.uiv.p2.eqcost-lin-vs-best-single.p-signflip\endcsname{$+$25.2}
\expandafter\def\csname val@rev.pol.perm.f20.uiv.p2.eqcost-lin-vs-best-single.p-signflip-lt\endcsname{0.749}
\expandafter\def\csname cnt@rev.pol.perm.f20.uiv.p2.eqcost-lin-vs-best-single.p-signflip-lt\endcsname{1}
\expandafter\def\csname pct@rev.pol.perm.f20.uiv.p2.eqcost-lin-vs-best-single.p-signflip-lt\endcsname{74.9}
\expandafter\def\csname ppt@rev.pol.perm.f20.uiv.p2.eqcost-lin-vs-best-single.p-signflip-lt\endcsname{$+$74.9}
\expandafter\def\csname val@rev.pol.perm.f20.uiv.p2.eqcost-lin-vs-best-single.p-t\endcsname{0.236}
\expandafter\def\csname cnt@rev.pol.perm.f20.uiv.p2.eqcost-lin-vs-best-single.p-t\endcsname{0}
\expandafter\def\csname pct@rev.pol.perm.f20.uiv.p2.eqcost-lin-vs-best-single.p-t\endcsname{23.6}
\expandafter\def\csname ppt@rev.pol.perm.f20.uiv.p2.eqcost-lin-vs-best-single.p-t\endcsname{$+$23.6}
\expandafter\def\csname val@rev.pol.perm.f20.uiv.p2.eqcost-lin-vs-best-single.p-t-lt\endcsname{0.764}
\expandafter\def\csname cnt@rev.pol.perm.f20.uiv.p2.eqcost-lin-vs-best-single.p-t-lt\endcsname{1}
\expandafter\def\csname pct@rev.pol.perm.f20.uiv.p2.eqcost-lin-vs-best-single.p-t-lt\endcsname{76.4}
\expandafter\def\csname ppt@rev.pol.perm.f20.uiv.p2.eqcost-lin-vs-best-single.p-t-lt\endcsname{$+$76.4}
\expandafter\def\csname val@rev.pol.perm.f20.uiv.p2.eqcost-lin-vs-generic-ppi-source.p-signflip\endcsname{0.338}
\expandafter\def\csname cnt@rev.pol.perm.f20.uiv.p2.eqcost-lin-vs-generic-ppi-source.p-signflip\endcsname{0}
\expandafter\def\csname pct@rev.pol.perm.f20.uiv.p2.eqcost-lin-vs-generic-ppi-source.p-signflip\endcsname{33.8}
\expandafter\def\csname ppt@rev.pol.perm.f20.uiv.p2.eqcost-lin-vs-generic-ppi-source.p-signflip\endcsname{$+$33.8}
\expandafter\def\csname val@rev.pol.perm.f20.uiv.p2.eqcost-lin-vs-generic-ppi-source.p-signflip-lt\endcsname{0.663}
\expandafter\def\csname cnt@rev.pol.perm.f20.uiv.p2.eqcost-lin-vs-generic-ppi-source.p-signflip-lt\endcsname{1}
\expandafter\def\csname pct@rev.pol.perm.f20.uiv.p2.eqcost-lin-vs-generic-ppi-source.p-signflip-lt\endcsname{66.3}
\expandafter\def\csname ppt@rev.pol.perm.f20.uiv.p2.eqcost-lin-vs-generic-ppi-source.p-signflip-lt\endcsname{$+$66.3}
\expandafter\def\csname val@rev.pol.perm.f20.uiv.p2.eqcost-lin-vs-generic-ppi-source.p-t\endcsname{0.330}
\expandafter\def\csname cnt@rev.pol.perm.f20.uiv.p2.eqcost-lin-vs-generic-ppi-source.p-t\endcsname{0}
\expandafter\def\csname pct@rev.pol.perm.f20.uiv.p2.eqcost-lin-vs-generic-ppi-source.p-t\endcsname{33.0}
\expandafter\def\csname ppt@rev.pol.perm.f20.uiv.p2.eqcost-lin-vs-generic-ppi-source.p-t\endcsname{$+$33.0}
\expandafter\def\csname val@rev.pol.perm.f20.uiv.p2.eqcost-lin-vs-generic-ppi-source.p-t-lt\endcsname{0.670}
\expandafter\def\csname cnt@rev.pol.perm.f20.uiv.p2.eqcost-lin-vs-generic-ppi-source.p-t-lt\endcsname{1}
\expandafter\def\csname pct@rev.pol.perm.f20.uiv.p2.eqcost-lin-vs-generic-ppi-source.p-t-lt\endcsname{67.0}
\expandafter\def\csname ppt@rev.pol.perm.f20.uiv.p2.eqcost-lin-vs-generic-ppi-source.p-t-lt\endcsname{$+$67.0}
\expandafter\def\csname val@rev.pol.perm.f20.uiv.p2.eqcost-lin-vs-replay-ppi-uniform.p-signflip\endcsname{0.030}
\expandafter\def\csname cnt@rev.pol.perm.f20.uiv.p2.eqcost-lin-vs-replay-ppi-uniform.p-signflip\endcsname{0}
\expandafter\def\csname pct@rev.pol.perm.f20.uiv.p2.eqcost-lin-vs-replay-ppi-uniform.p-signflip\endcsname{3.0}
\expandafter\def\csname ppt@rev.pol.perm.f20.uiv.p2.eqcost-lin-vs-replay-ppi-uniform.p-signflip\endcsname{$+$3.0}
\expandafter\def\csname val@rev.pol.perm.f20.uiv.p2.eqcost-lin-vs-replay-ppi-uniform.p-signflip-lt\endcsname{0.972}
\expandafter\def\csname cnt@rev.pol.perm.f20.uiv.p2.eqcost-lin-vs-replay-ppi-uniform.p-signflip-lt\endcsname{1}
\expandafter\def\csname pct@rev.pol.perm.f20.uiv.p2.eqcost-lin-vs-replay-ppi-uniform.p-signflip-lt\endcsname{97.2}
\expandafter\def\csname ppt@rev.pol.perm.f20.uiv.p2.eqcost-lin-vs-replay-ppi-uniform.p-signflip-lt\endcsname{$+$97.2}
\expandafter\def\csname val@rev.pol.perm.f20.uiv.p2.eqcost-lin-vs-replay-ppi-uniform.p-t\endcsname{0.029}
\expandafter\def\csname cnt@rev.pol.perm.f20.uiv.p2.eqcost-lin-vs-replay-ppi-uniform.p-t\endcsname{0}
\expandafter\def\csname pct@rev.pol.perm.f20.uiv.p2.eqcost-lin-vs-replay-ppi-uniform.p-t\endcsname{2.9}
\expandafter\def\csname ppt@rev.pol.perm.f20.uiv.p2.eqcost-lin-vs-replay-ppi-uniform.p-t\endcsname{$+$2.9}
\expandafter\def\csname val@rev.pol.perm.f20.uiv.p2.eqcost-lin-vs-replay-ppi-uniform.p-t-lt\endcsname{0.971}
\expandafter\def\csname cnt@rev.pol.perm.f20.uiv.p2.eqcost-lin-vs-replay-ppi-uniform.p-t-lt\endcsname{1}
\expandafter\def\csname pct@rev.pol.perm.f20.uiv.p2.eqcost-lin-vs-replay-ppi-uniform.p-t-lt\endcsname{97.1}
\expandafter\def\csname ppt@rev.pol.perm.f20.uiv.p2.eqcost-lin-vs-replay-ppi-uniform.p-t-lt\endcsname{$+$97.1}
\expandafter\def\csname val@rev.pol.perm.f20.uiv.p2.eqcost-lin-vs-uniform-fresh.p-signflip\endcsname{0.519}
\expandafter\def\csname cnt@rev.pol.perm.f20.uiv.p2.eqcost-lin-vs-uniform-fresh.p-signflip\endcsname{1}
\expandafter\def\csname pct@rev.pol.perm.f20.uiv.p2.eqcost-lin-vs-uniform-fresh.p-signflip\endcsname{51.9}
\expandafter\def\csname ppt@rev.pol.perm.f20.uiv.p2.eqcost-lin-vs-uniform-fresh.p-signflip\endcsname{$+$51.9}
\expandafter\def\csname val@rev.pol.perm.f20.uiv.p2.eqcost-lin-vs-uniform-fresh.p-signflip-lt\endcsname{0.482}
\expandafter\def\csname cnt@rev.pol.perm.f20.uiv.p2.eqcost-lin-vs-uniform-fresh.p-signflip-lt\endcsname{0}
\expandafter\def\csname pct@rev.pol.perm.f20.uiv.p2.eqcost-lin-vs-uniform-fresh.p-signflip-lt\endcsname{48.2}
\expandafter\def\csname ppt@rev.pol.perm.f20.uiv.p2.eqcost-lin-vs-uniform-fresh.p-signflip-lt\endcsname{$+$48.2}
\expandafter\def\csname val@rev.pol.perm.f20.uiv.p2.eqcost-lin-vs-uniform-fresh.p-t\endcsname{0.525}
\expandafter\def\csname cnt@rev.pol.perm.f20.uiv.p2.eqcost-lin-vs-uniform-fresh.p-t\endcsname{1}
\expandafter\def\csname pct@rev.pol.perm.f20.uiv.p2.eqcost-lin-vs-uniform-fresh.p-t\endcsname{52.5}
\expandafter\def\csname ppt@rev.pol.perm.f20.uiv.p2.eqcost-lin-vs-uniform-fresh.p-t\endcsname{$+$52.5}
\expandafter\def\csname val@rev.pol.perm.f20.uiv.p2.eqcost-lin-vs-uniform-fresh.p-t-lt\endcsname{0.475}
\expandafter\def\csname cnt@rev.pol.perm.f20.uiv.p2.eqcost-lin-vs-uniform-fresh.p-t-lt\endcsname{0}
\expandafter\def\csname pct@rev.pol.perm.f20.uiv.p2.eqcost-lin-vs-uniform-fresh.p-t-lt\endcsname{47.5}
\expandafter\def\csname ppt@rev.pol.perm.f20.uiv.p2.eqcost-lin-vs-uniform-fresh.p-t-lt\endcsname{$+$47.5}
\expandafter\def\csname val@rev.pol.perm.f20.uiv.p2.rcr-vs-full-recert-ni.p-signflip\endcsname{0.024}
\expandafter\def\csname cnt@rev.pol.perm.f20.uiv.p2.rcr-vs-full-recert-ni.p-signflip\endcsname{0}
\expandafter\def\csname pct@rev.pol.perm.f20.uiv.p2.rcr-vs-full-recert-ni.p-signflip\endcsname{2.4}
\expandafter\def\csname ppt@rev.pol.perm.f20.uiv.p2.rcr-vs-full-recert-ni.p-signflip\endcsname{$+$2.4}
\expandafter\def\csname val@rev.pol.perm.f20.uiv.p2.rcr-vs-full-recert-ni.p-t\endcsname{0.029}
\expandafter\def\csname cnt@rev.pol.perm.f20.uiv.p2.rcr-vs-full-recert-ni.p-t\endcsname{0}
\expandafter\def\csname pct@rev.pol.perm.f20.uiv.p2.rcr-vs-full-recert-ni.p-t\endcsname{2.9}
\expandafter\def\csname ppt@rev.pol.perm.f20.uiv.p2.rcr-vs-full-recert-ni.p-t\endcsname{$+$2.9}
\expandafter\def\csname val@rev.pol.perm.f20.uiv.p2.rcr-vs-generic-ppi-source.p-signflip\endcsname{0.029}
\expandafter\def\csname cnt@rev.pol.perm.f20.uiv.p2.rcr-vs-generic-ppi-source.p-signflip\endcsname{0}
\expandafter\def\csname pct@rev.pol.perm.f20.uiv.p2.rcr-vs-generic-ppi-source.p-signflip\endcsname{2.9}
\expandafter\def\csname ppt@rev.pol.perm.f20.uiv.p2.rcr-vs-generic-ppi-source.p-signflip\endcsname{$+$2.9}
\expandafter\def\csname val@rev.pol.perm.f20.uiv.p2.rcr-vs-generic-ppi-source.p-signflip-holm\endcsname{0.145}
\expandafter\def\csname cnt@rev.pol.perm.f20.uiv.p2.rcr-vs-generic-ppi-source.p-signflip-holm\endcsname{0}
\expandafter\def\csname pct@rev.pol.perm.f20.uiv.p2.rcr-vs-generic-ppi-source.p-signflip-holm\endcsname{14.5}
\expandafter\def\csname ppt@rev.pol.perm.f20.uiv.p2.rcr-vs-generic-ppi-source.p-signflip-holm\endcsname{$+$14.5}
\expandafter\def\csname val@rev.pol.perm.f20.uiv.p2.rcr-vs-generic-ppi-source.p-t\endcsname{0.027}
\expandafter\def\csname cnt@rev.pol.perm.f20.uiv.p2.rcr-vs-generic-ppi-source.p-t\endcsname{0}
\expandafter\def\csname pct@rev.pol.perm.f20.uiv.p2.rcr-vs-generic-ppi-source.p-t\endcsname{2.7}
\expandafter\def\csname ppt@rev.pol.perm.f20.uiv.p2.rcr-vs-generic-ppi-source.p-t\endcsname{$+$2.7}
\expandafter\def\csname val@rev.pol.perm.f20.uiv.p2.rcr-vs-generic-ppi-source.p-t-holm\endcsname{0.134}
\expandafter\def\csname cnt@rev.pol.perm.f20.uiv.p2.rcr-vs-generic-ppi-source.p-t-holm\endcsname{0}
\expandafter\def\csname pct@rev.pol.perm.f20.uiv.p2.rcr-vs-generic-ppi-source.p-t-holm\endcsname{13.4}
\expandafter\def\csname ppt@rev.pol.perm.f20.uiv.p2.rcr-vs-generic-ppi-source.p-t-holm\endcsname{$+$13.4}
\expandafter\def\csname val@rev.pol.perm.f20.uiv.p2.rcr-vs-replay-ppi-uniform.p-signflip\endcsname{0.038}
\expandafter\def\csname cnt@rev.pol.perm.f20.uiv.p2.rcr-vs-replay-ppi-uniform.p-signflip\endcsname{0}
\expandafter\def\csname pct@rev.pol.perm.f20.uiv.p2.rcr-vs-replay-ppi-uniform.p-signflip\endcsname{3.8}
\expandafter\def\csname ppt@rev.pol.perm.f20.uiv.p2.rcr-vs-replay-ppi-uniform.p-signflip\endcsname{$+$3.8}
\expandafter\def\csname val@rev.pol.perm.f20.uiv.p2.rcr-vs-replay-ppi-uniform.p-signflip-holm\endcsname{0.145}
\expandafter\def\csname cnt@rev.pol.perm.f20.uiv.p2.rcr-vs-replay-ppi-uniform.p-signflip-holm\endcsname{0}
\expandafter\def\csname pct@rev.pol.perm.f20.uiv.p2.rcr-vs-replay-ppi-uniform.p-signflip-holm\endcsname{14.5}
\expandafter\def\csname ppt@rev.pol.perm.f20.uiv.p2.rcr-vs-replay-ppi-uniform.p-signflip-holm\endcsname{$+$14.5}
\expandafter\def\csname val@rev.pol.perm.f20.uiv.p2.rcr-vs-replay-ppi-uniform.p-t\endcsname{0.029}
\expandafter\def\csname cnt@rev.pol.perm.f20.uiv.p2.rcr-vs-replay-ppi-uniform.p-t\endcsname{0}
\expandafter\def\csname pct@rev.pol.perm.f20.uiv.p2.rcr-vs-replay-ppi-uniform.p-t\endcsname{2.9}
\expandafter\def\csname ppt@rev.pol.perm.f20.uiv.p2.rcr-vs-replay-ppi-uniform.p-t\endcsname{$+$2.9}
\expandafter\def\csname val@rev.pol.perm.f20.uiv.p2.rcr-vs-replay-ppi-uniform.p-t-holm\endcsname{0.134}
\expandafter\def\csname cnt@rev.pol.perm.f20.uiv.p2.rcr-vs-replay-ppi-uniform.p-t-holm\endcsname{0}
\expandafter\def\csname pct@rev.pol.perm.f20.uiv.p2.rcr-vs-replay-ppi-uniform.p-t-holm\endcsname{13.4}
\expandafter\def\csname ppt@rev.pol.perm.f20.uiv.p2.rcr-vs-replay-ppi-uniform.p-t-holm\endcsname{$+$13.4}
\expandafter\def\csname val@rev.pol.perm.n.ni.n-pass-signflip\endcsname{3}
\expandafter\def\csname cnt@rev.pol.perm.n.ni.n-pass-signflip\endcsname{3}
\expandafter\def\csname pct@rev.pol.perm.n.ni.n-pass-signflip\endcsname{300.0}
\expandafter\def\csname ppt@rev.pol.perm.n.ni.n-pass-signflip\endcsname{$+$300.0}
\expandafter\def\csname val@rev.pol.perm.n.ni.n-pass-t\endcsname{3}
\expandafter\def\csname cnt@rev.pol.perm.n.ni.n-pass-t\endcsname{3}
\expandafter\def\csname pct@rev.pol.perm.n.ni.n-pass-t\endcsname{300.0}
\expandafter\def\csname ppt@rev.pol.perm.n.ni.n-pass-t\endcsname{$+$300.0}
\expandafter\def\csname val@rev.pol.perm.n.signflip-gt.n-rejected\endcsname{2}
\expandafter\def\csname cnt@rev.pol.perm.n.signflip-gt.n-rejected\endcsname{2}
\expandafter\def\csname pct@rev.pol.perm.n.signflip-gt.n-rejected\endcsname{200.0}
\expandafter\def\csname ppt@rev.pol.perm.n.signflip-gt.n-rejected\endcsname{$+$200.0}
\expandafter\def\csname val@rev.pol.perm.n.sspro.p1.eqcost-lin-vs-best-single.p-signflip\endcsname{0.316}
\expandafter\def\csname cnt@rev.pol.perm.n.sspro.p1.eqcost-lin-vs-best-single.p-signflip\endcsname{0}
\expandafter\def\csname pct@rev.pol.perm.n.sspro.p1.eqcost-lin-vs-best-single.p-signflip\endcsname{31.6}
\expandafter\def\csname ppt@rev.pol.perm.n.sspro.p1.eqcost-lin-vs-best-single.p-signflip\endcsname{$+$31.6}
\expandafter\def\csname val@rev.pol.perm.n.sspro.p1.eqcost-lin-vs-best-single.p-signflip-lt\endcsname{0.685}
\expandafter\def\csname cnt@rev.pol.perm.n.sspro.p1.eqcost-lin-vs-best-single.p-signflip-lt\endcsname{1}
\expandafter\def\csname pct@rev.pol.perm.n.sspro.p1.eqcost-lin-vs-best-single.p-signflip-lt\endcsname{68.5}
\expandafter\def\csname ppt@rev.pol.perm.n.sspro.p1.eqcost-lin-vs-best-single.p-signflip-lt\endcsname{$+$68.5}
\expandafter\def\csname val@rev.pol.perm.n.sspro.p1.eqcost-lin-vs-best-single.p-t\endcsname{0.309}
\expandafter\def\csname cnt@rev.pol.perm.n.sspro.p1.eqcost-lin-vs-best-single.p-t\endcsname{0}
\expandafter\def\csname pct@rev.pol.perm.n.sspro.p1.eqcost-lin-vs-best-single.p-t\endcsname{30.9}
\expandafter\def\csname ppt@rev.pol.perm.n.sspro.p1.eqcost-lin-vs-best-single.p-t\endcsname{$+$30.9}
\expandafter\def\csname val@rev.pol.perm.n.sspro.p1.eqcost-lin-vs-best-single.p-t-lt\endcsname{0.691}
\expandafter\def\csname cnt@rev.pol.perm.n.sspro.p1.eqcost-lin-vs-best-single.p-t-lt\endcsname{1}
\expandafter\def\csname pct@rev.pol.perm.n.sspro.p1.eqcost-lin-vs-best-single.p-t-lt\endcsname{69.1}
\expandafter\def\csname ppt@rev.pol.perm.n.sspro.p1.eqcost-lin-vs-best-single.p-t-lt\endcsname{$+$69.1}
\expandafter\def\csname val@rev.pol.perm.n.sspro.p1.eqcost-lin-vs-generic-ppi-source.p-signflip\endcsname{0.951}
\expandafter\def\csname cnt@rev.pol.perm.n.sspro.p1.eqcost-lin-vs-generic-ppi-source.p-signflip\endcsname{1}
\expandafter\def\csname pct@rev.pol.perm.n.sspro.p1.eqcost-lin-vs-generic-ppi-source.p-signflip\endcsname{95.1}
\expandafter\def\csname ppt@rev.pol.perm.n.sspro.p1.eqcost-lin-vs-generic-ppi-source.p-signflip\endcsname{$+$95.1}
\expandafter\def\csname val@rev.pol.perm.n.sspro.p1.eqcost-lin-vs-generic-ppi-source.p-signflip-lt\endcsname{0.052}
\expandafter\def\csname cnt@rev.pol.perm.n.sspro.p1.eqcost-lin-vs-generic-ppi-source.p-signflip-lt\endcsname{0}
\expandafter\def\csname pct@rev.pol.perm.n.sspro.p1.eqcost-lin-vs-generic-ppi-source.p-signflip-lt\endcsname{5.2}
\expandafter\def\csname ppt@rev.pol.perm.n.sspro.p1.eqcost-lin-vs-generic-ppi-source.p-signflip-lt\endcsname{$+$5.2}
\expandafter\def\csname val@rev.pol.perm.n.sspro.p1.eqcost-lin-vs-generic-ppi-source.p-t\endcsname{0.956}
\expandafter\def\csname cnt@rev.pol.perm.n.sspro.p1.eqcost-lin-vs-generic-ppi-source.p-t\endcsname{1}
\expandafter\def\csname pct@rev.pol.perm.n.sspro.p1.eqcost-lin-vs-generic-ppi-source.p-t\endcsname{95.6}
\expandafter\def\csname ppt@rev.pol.perm.n.sspro.p1.eqcost-lin-vs-generic-ppi-source.p-t\endcsname{$+$95.6}
\expandafter\def\csname val@rev.pol.perm.n.sspro.p1.eqcost-lin-vs-generic-ppi-source.p-t-lt\endcsname{0.044}
\expandafter\def\csname cnt@rev.pol.perm.n.sspro.p1.eqcost-lin-vs-generic-ppi-source.p-t-lt\endcsname{0}
\expandafter\def\csname pct@rev.pol.perm.n.sspro.p1.eqcost-lin-vs-generic-ppi-source.p-t-lt\endcsname{4.4}
\expandafter\def\csname ppt@rev.pol.perm.n.sspro.p1.eqcost-lin-vs-generic-ppi-source.p-t-lt\endcsname{$+$4.4}
\expandafter\def\csname val@rev.pol.perm.n.sspro.p1.eqcost-lin-vs-replay-ppi-uniform.p-signflip\endcsname{0.032}
\expandafter\def\csname cnt@rev.pol.perm.n.sspro.p1.eqcost-lin-vs-replay-ppi-uniform.p-signflip\endcsname{0}
\expandafter\def\csname pct@rev.pol.perm.n.sspro.p1.eqcost-lin-vs-replay-ppi-uniform.p-signflip\endcsname{3.2}
\expandafter\def\csname ppt@rev.pol.perm.n.sspro.p1.eqcost-lin-vs-replay-ppi-uniform.p-signflip\endcsname{$+$3.2}
\expandafter\def\csname val@rev.pol.perm.n.sspro.p1.eqcost-lin-vs-replay-ppi-uniform.p-signflip-lt\endcsname{0.971}
\expandafter\def\csname cnt@rev.pol.perm.n.sspro.p1.eqcost-lin-vs-replay-ppi-uniform.p-signflip-lt\endcsname{1}
\expandafter\def\csname pct@rev.pol.perm.n.sspro.p1.eqcost-lin-vs-replay-ppi-uniform.p-signflip-lt\endcsname{97.1}
\expandafter\def\csname ppt@rev.pol.perm.n.sspro.p1.eqcost-lin-vs-replay-ppi-uniform.p-signflip-lt\endcsname{$+$97.1}
\expandafter\def\csname val@rev.pol.perm.n.sspro.p1.eqcost-lin-vs-replay-ppi-uniform.p-t\endcsname{0.043}
\expandafter\def\csname cnt@rev.pol.perm.n.sspro.p1.eqcost-lin-vs-replay-ppi-uniform.p-t\endcsname{0}
\expandafter\def\csname pct@rev.pol.perm.n.sspro.p1.eqcost-lin-vs-replay-ppi-uniform.p-t\endcsname{4.3}
\expandafter\def\csname ppt@rev.pol.perm.n.sspro.p1.eqcost-lin-vs-replay-ppi-uniform.p-t\endcsname{$+$4.3}
\expandafter\def\csname val@rev.pol.perm.n.sspro.p1.eqcost-lin-vs-replay-ppi-uniform.p-t-lt\endcsname{0.957}
\expandafter\def\csname cnt@rev.pol.perm.n.sspro.p1.eqcost-lin-vs-replay-ppi-uniform.p-t-lt\endcsname{1}
\expandafter\def\csname pct@rev.pol.perm.n.sspro.p1.eqcost-lin-vs-replay-ppi-uniform.p-t-lt\endcsname{95.7}
\expandafter\def\csname ppt@rev.pol.perm.n.sspro.p1.eqcost-lin-vs-replay-ppi-uniform.p-t-lt\endcsname{$+$95.7}
\expandafter\def\csname val@rev.pol.perm.n.sspro.p1.eqcost-lin-vs-uniform-fresh.p-signflip\endcsname{0.954}
\expandafter\def\csname cnt@rev.pol.perm.n.sspro.p1.eqcost-lin-vs-uniform-fresh.p-signflip\endcsname{1}
\expandafter\def\csname pct@rev.pol.perm.n.sspro.p1.eqcost-lin-vs-uniform-fresh.p-signflip\endcsname{95.4}
\expandafter\def\csname ppt@rev.pol.perm.n.sspro.p1.eqcost-lin-vs-uniform-fresh.p-signflip\endcsname{$+$95.4}
\expandafter\def\csname val@rev.pol.perm.n.sspro.p1.eqcost-lin-vs-uniform-fresh.p-signflip-lt\endcsname{0.047}
\expandafter\def\csname cnt@rev.pol.perm.n.sspro.p1.eqcost-lin-vs-uniform-fresh.p-signflip-lt\endcsname{0}
\expandafter\def\csname pct@rev.pol.perm.n.sspro.p1.eqcost-lin-vs-uniform-fresh.p-signflip-lt\endcsname{4.7}
\expandafter\def\csname ppt@rev.pol.perm.n.sspro.p1.eqcost-lin-vs-uniform-fresh.p-signflip-lt\endcsname{$+$4.7}
\expandafter\def\csname val@rev.pol.perm.n.sspro.p1.eqcost-lin-vs-uniform-fresh.p-t\endcsname{0.957}
\expandafter\def\csname cnt@rev.pol.perm.n.sspro.p1.eqcost-lin-vs-uniform-fresh.p-t\endcsname{1}
\expandafter\def\csname pct@rev.pol.perm.n.sspro.p1.eqcost-lin-vs-uniform-fresh.p-t\endcsname{95.7}
\expandafter\def\csname ppt@rev.pol.perm.n.sspro.p1.eqcost-lin-vs-uniform-fresh.p-t\endcsname{$+$95.7}
\expandafter\def\csname val@rev.pol.perm.n.sspro.p1.eqcost-lin-vs-uniform-fresh.p-t-lt\endcsname{0.043}
\expandafter\def\csname cnt@rev.pol.perm.n.sspro.p1.eqcost-lin-vs-uniform-fresh.p-t-lt\endcsname{0}
\expandafter\def\csname pct@rev.pol.perm.n.sspro.p1.eqcost-lin-vs-uniform-fresh.p-t-lt\endcsname{4.3}
\expandafter\def\csname ppt@rev.pol.perm.n.sspro.p1.eqcost-lin-vs-uniform-fresh.p-t-lt\endcsname{$+$4.3}
\expandafter\def\csname val@rev.pol.perm.n.sspro.p1.rcr-vs-full-recert-ni.p-signflip\endcsname{0.006}
\expandafter\def\csname cnt@rev.pol.perm.n.sspro.p1.rcr-vs-full-recert-ni.p-signflip\endcsname{0}
\expandafter\def\csname pct@rev.pol.perm.n.sspro.p1.rcr-vs-full-recert-ni.p-signflip\endcsname{0.6}
\expandafter\def\csname ppt@rev.pol.perm.n.sspro.p1.rcr-vs-full-recert-ni.p-signflip\endcsname{$+$0.6}
\expandafter\def\csname val@rev.pol.perm.n.sspro.p1.rcr-vs-full-recert-ni.p-t\endcsname{0.004}
\expandafter\def\csname cnt@rev.pol.perm.n.sspro.p1.rcr-vs-full-recert-ni.p-t\endcsname{0}
\expandafter\def\csname pct@rev.pol.perm.n.sspro.p1.rcr-vs-full-recert-ni.p-t\endcsname{0.4}
\expandafter\def\csname ppt@rev.pol.perm.n.sspro.p1.rcr-vs-full-recert-ni.p-t\endcsname{$+$0.4}
\expandafter\def\csname val@rev.pol.perm.n.sspro.p1.rcr-vs-generic-ppi-source.p-signflip\endcsname{0.020}
\expandafter\def\csname cnt@rev.pol.perm.n.sspro.p1.rcr-vs-generic-ppi-source.p-signflip\endcsname{0}
\expandafter\def\csname pct@rev.pol.perm.n.sspro.p1.rcr-vs-generic-ppi-source.p-signflip\endcsname{2.0}
\expandafter\def\csname ppt@rev.pol.perm.n.sspro.p1.rcr-vs-generic-ppi-source.p-signflip\endcsname{$+$2.0}
\expandafter\def\csname val@rev.pol.perm.n.sspro.p1.rcr-vs-generic-ppi-source.p-signflip-holm\endcsname{0.100}
\expandafter\def\csname cnt@rev.pol.perm.n.sspro.p1.rcr-vs-generic-ppi-source.p-signflip-holm\endcsname{0}
\expandafter\def\csname pct@rev.pol.perm.n.sspro.p1.rcr-vs-generic-ppi-source.p-signflip-holm\endcsname{10.0}
\expandafter\def\csname ppt@rev.pol.perm.n.sspro.p1.rcr-vs-generic-ppi-source.p-signflip-holm\endcsname{$+$10.0}
\expandafter\def\csname val@rev.pol.perm.n.sspro.p1.rcr-vs-generic-ppi-source.p-t\endcsname{0.046}
\expandafter\def\csname cnt@rev.pol.perm.n.sspro.p1.rcr-vs-generic-ppi-source.p-t\endcsname{0}
\expandafter\def\csname pct@rev.pol.perm.n.sspro.p1.rcr-vs-generic-ppi-source.p-t\endcsname{4.6}
\expandafter\def\csname ppt@rev.pol.perm.n.sspro.p1.rcr-vs-generic-ppi-source.p-t\endcsname{$+$4.6}
\expandafter\def\csname val@rev.pol.perm.n.sspro.p1.rcr-vs-generic-ppi-source.p-t-holm\endcsname{0.214}
\expandafter\def\csname cnt@rev.pol.perm.n.sspro.p1.rcr-vs-generic-ppi-source.p-t-holm\endcsname{0}
\expandafter\def\csname pct@rev.pol.perm.n.sspro.p1.rcr-vs-generic-ppi-source.p-t-holm\endcsname{21.4}
\expandafter\def\csname ppt@rev.pol.perm.n.sspro.p1.rcr-vs-generic-ppi-source.p-t-holm\endcsname{$+$21.4}
\expandafter\def\csname val@rev.pol.perm.n.sspro.p1.rcr-vs-replay-ppi-uniform.p-signflip\endcsname{0.032}
\expandafter\def\csname cnt@rev.pol.perm.n.sspro.p1.rcr-vs-replay-ppi-uniform.p-signflip\endcsname{0}
\expandafter\def\csname pct@rev.pol.perm.n.sspro.p1.rcr-vs-replay-ppi-uniform.p-signflip\endcsname{3.2}
\expandafter\def\csname ppt@rev.pol.perm.n.sspro.p1.rcr-vs-replay-ppi-uniform.p-signflip\endcsname{$+$3.2}
\expandafter\def\csname val@rev.pol.perm.n.sspro.p1.rcr-vs-replay-ppi-uniform.p-signflip-holm\endcsname{0.129}
\expandafter\def\csname cnt@rev.pol.perm.n.sspro.p1.rcr-vs-replay-ppi-uniform.p-signflip-holm\endcsname{0}
\expandafter\def\csname pct@rev.pol.perm.n.sspro.p1.rcr-vs-replay-ppi-uniform.p-signflip-holm\endcsname{12.9}
\expandafter\def\csname ppt@rev.pol.perm.n.sspro.p1.rcr-vs-replay-ppi-uniform.p-signflip-holm\endcsname{$+$12.9}
\expandafter\def\csname val@rev.pol.perm.n.sspro.p1.rcr-vs-replay-ppi-uniform.p-t\endcsname{0.043}
\expandafter\def\csname cnt@rev.pol.perm.n.sspro.p1.rcr-vs-replay-ppi-uniform.p-t\endcsname{0}
\expandafter\def\csname pct@rev.pol.perm.n.sspro.p1.rcr-vs-replay-ppi-uniform.p-t\endcsname{4.3}
\expandafter\def\csname ppt@rev.pol.perm.n.sspro.p1.rcr-vs-replay-ppi-uniform.p-t\endcsname{$+$4.3}
\expandafter\def\csname val@rev.pol.perm.n.sspro.p1.rcr-vs-replay-ppi-uniform.p-t-holm\endcsname{0.214}
\expandafter\def\csname cnt@rev.pol.perm.n.sspro.p1.rcr-vs-replay-ppi-uniform.p-t-holm\endcsname{0}
\expandafter\def\csname pct@rev.pol.perm.n.sspro.p1.rcr-vs-replay-ppi-uniform.p-t-holm\endcsname{21.4}
\expandafter\def\csname ppt@rev.pol.perm.n.sspro.p1.rcr-vs-replay-ppi-uniform.p-t-holm\endcsname{$+$21.4}
\expandafter\def\csname val@rev.pol.perm.n.sspro.p2.eqcost-lin-vs-best-single.p-signflip\endcsname{0.995}
\expandafter\def\csname cnt@rev.pol.perm.n.sspro.p2.eqcost-lin-vs-best-single.p-signflip\endcsname{1}
\expandafter\def\csname pct@rev.pol.perm.n.sspro.p2.eqcost-lin-vs-best-single.p-signflip\endcsname{99.5}
\expandafter\def\csname ppt@rev.pol.perm.n.sspro.p2.eqcost-lin-vs-best-single.p-signflip\endcsname{$+$99.5}
\expandafter\def\csname val@rev.pol.perm.n.sspro.p2.eqcost-lin-vs-best-single.p-signflip-lt\endcsname{0.006}
\expandafter\def\csname cnt@rev.pol.perm.n.sspro.p2.eqcost-lin-vs-best-single.p-signflip-lt\endcsname{0}
\expandafter\def\csname pct@rev.pol.perm.n.sspro.p2.eqcost-lin-vs-best-single.p-signflip-lt\endcsname{0.6}
\expandafter\def\csname ppt@rev.pol.perm.n.sspro.p2.eqcost-lin-vs-best-single.p-signflip-lt\endcsname{$+$0.6}
\expandafter\def\csname val@rev.pol.perm.n.sspro.p2.eqcost-lin-vs-best-single.p-t\endcsname{0.994}
\expandafter\def\csname cnt@rev.pol.perm.n.sspro.p2.eqcost-lin-vs-best-single.p-t\endcsname{1}
\expandafter\def\csname pct@rev.pol.perm.n.sspro.p2.eqcost-lin-vs-best-single.p-t\endcsname{99.4}
\expandafter\def\csname ppt@rev.pol.perm.n.sspro.p2.eqcost-lin-vs-best-single.p-t\endcsname{$+$99.4}
\expandafter\def\csname val@rev.pol.perm.n.sspro.p2.eqcost-lin-vs-best-single.p-t-lt\endcsname{0.006}
\expandafter\def\csname cnt@rev.pol.perm.n.sspro.p2.eqcost-lin-vs-best-single.p-t-lt\endcsname{0}
\expandafter\def\csname pct@rev.pol.perm.n.sspro.p2.eqcost-lin-vs-best-single.p-t-lt\endcsname{0.6}
\expandafter\def\csname ppt@rev.pol.perm.n.sspro.p2.eqcost-lin-vs-best-single.p-t-lt\endcsname{$+$0.6}
\expandafter\def\csname val@rev.pol.perm.n.sspro.p2.eqcost-lin-vs-generic-ppi-source.p-signflip\endcsname{1.000}
\expandafter\def\csname cnt@rev.pol.perm.n.sspro.p2.eqcost-lin-vs-generic-ppi-source.p-signflip\endcsname{1}
\expandafter\def\csname pct@rev.pol.perm.n.sspro.p2.eqcost-lin-vs-generic-ppi-source.p-signflip\endcsname{100.0}
\expandafter\def\csname ppt@rev.pol.perm.n.sspro.p2.eqcost-lin-vs-generic-ppi-source.p-signflip\endcsname{$+$100.0}
\expandafter\def\csname val@rev.pol.perm.n.sspro.p2.eqcost-lin-vs-generic-ppi-source.p-signflip-lt\endcsname{0.001}
\expandafter\def\csname cnt@rev.pol.perm.n.sspro.p2.eqcost-lin-vs-generic-ppi-source.p-signflip-lt\endcsname{0}
\expandafter\def\csname pct@rev.pol.perm.n.sspro.p2.eqcost-lin-vs-generic-ppi-source.p-signflip-lt\endcsname{0.1}
\expandafter\def\csname ppt@rev.pol.perm.n.sspro.p2.eqcost-lin-vs-generic-ppi-source.p-signflip-lt\endcsname{$+$0.1}
\expandafter\def\csname val@rev.pol.perm.n.sspro.p2.eqcost-lin-vs-generic-ppi-source.p-t\endcsname{0.999}
\expandafter\def\csname cnt@rev.pol.perm.n.sspro.p2.eqcost-lin-vs-generic-ppi-source.p-t\endcsname{1}
\expandafter\def\csname pct@rev.pol.perm.n.sspro.p2.eqcost-lin-vs-generic-ppi-source.p-t\endcsname{99.9}
\expandafter\def\csname ppt@rev.pol.perm.n.sspro.p2.eqcost-lin-vs-generic-ppi-source.p-t\endcsname{$+$99.9}
\expandafter\def\csname val@rev.pol.perm.n.sspro.p2.eqcost-lin-vs-generic-ppi-source.p-t-lt\endcsname{0.001}
\expandafter\def\csname cnt@rev.pol.perm.n.sspro.p2.eqcost-lin-vs-generic-ppi-source.p-t-lt\endcsname{0}
\expandafter\def\csname pct@rev.pol.perm.n.sspro.p2.eqcost-lin-vs-generic-ppi-source.p-t-lt\endcsname{0.1}
\expandafter\def\csname ppt@rev.pol.perm.n.sspro.p2.eqcost-lin-vs-generic-ppi-source.p-t-lt\endcsname{$+$0.1}
\expandafter\def\csname val@rev.pol.perm.n.sspro.p2.eqcost-lin-vs-replay-ppi-uniform.p-signflip\endcsname{0.500}
\expandafter\def\csname cnt@rev.pol.perm.n.sspro.p2.eqcost-lin-vs-replay-ppi-uniform.p-signflip\endcsname{0}
\expandafter\def\csname pct@rev.pol.perm.n.sspro.p2.eqcost-lin-vs-replay-ppi-uniform.p-signflip\endcsname{50.0}
\expandafter\def\csname ppt@rev.pol.perm.n.sspro.p2.eqcost-lin-vs-replay-ppi-uniform.p-signflip\endcsname{$+$50.0}
\expandafter\def\csname val@rev.pol.perm.n.sspro.p2.eqcost-lin-vs-replay-ppi-uniform.p-signflip-lt\endcsname{0.501}
\expandafter\def\csname cnt@rev.pol.perm.n.sspro.p2.eqcost-lin-vs-replay-ppi-uniform.p-signflip-lt\endcsname{1}
\expandafter\def\csname pct@rev.pol.perm.n.sspro.p2.eqcost-lin-vs-replay-ppi-uniform.p-signflip-lt\endcsname{50.1}
\expandafter\def\csname ppt@rev.pol.perm.n.sspro.p2.eqcost-lin-vs-replay-ppi-uniform.p-signflip-lt\endcsname{$+$50.1}
\expandafter\def\csname val@rev.pol.perm.n.sspro.p2.eqcost-lin-vs-replay-ppi-uniform.p-t\endcsname{0.471}
\expandafter\def\csname cnt@rev.pol.perm.n.sspro.p2.eqcost-lin-vs-replay-ppi-uniform.p-t\endcsname{0}
\expandafter\def\csname pct@rev.pol.perm.n.sspro.p2.eqcost-lin-vs-replay-ppi-uniform.p-t\endcsname{47.1}
\expandafter\def\csname ppt@rev.pol.perm.n.sspro.p2.eqcost-lin-vs-replay-ppi-uniform.p-t\endcsname{$+$47.1}
\expandafter\def\csname val@rev.pol.perm.n.sspro.p2.eqcost-lin-vs-replay-ppi-uniform.p-t-lt\endcsname{0.529}
\expandafter\def\csname cnt@rev.pol.perm.n.sspro.p2.eqcost-lin-vs-replay-ppi-uniform.p-t-lt\endcsname{1}
\expandafter\def\csname pct@rev.pol.perm.n.sspro.p2.eqcost-lin-vs-replay-ppi-uniform.p-t-lt\endcsname{52.9}
\expandafter\def\csname ppt@rev.pol.perm.n.sspro.p2.eqcost-lin-vs-replay-ppi-uniform.p-t-lt\endcsname{$+$52.9}
\expandafter\def\csname val@rev.pol.perm.n.sspro.p2.eqcost-lin-vs-uniform-fresh.p-signflip\endcsname{1.000}
\expandafter\def\csname cnt@rev.pol.perm.n.sspro.p2.eqcost-lin-vs-uniform-fresh.p-signflip\endcsname{1}
\expandafter\def\csname pct@rev.pol.perm.n.sspro.p2.eqcost-lin-vs-uniform-fresh.p-signflip\endcsname{100.0}
\expandafter\def\csname ppt@rev.pol.perm.n.sspro.p2.eqcost-lin-vs-uniform-fresh.p-signflip\endcsname{$+$100.0}
\expandafter\def\csname val@rev.pol.perm.n.sspro.p2.eqcost-lin-vs-uniform-fresh.p-signflip-lt\endcsname{0.001}
\expandafter\def\csname cnt@rev.pol.perm.n.sspro.p2.eqcost-lin-vs-uniform-fresh.p-signflip-lt\endcsname{0}
\expandafter\def\csname pct@rev.pol.perm.n.sspro.p2.eqcost-lin-vs-uniform-fresh.p-signflip-lt\endcsname{0.1}
\expandafter\def\csname ppt@rev.pol.perm.n.sspro.p2.eqcost-lin-vs-uniform-fresh.p-signflip-lt\endcsname{$+$0.1}
\expandafter\def\csname val@rev.pol.perm.n.sspro.p2.eqcost-lin-vs-uniform-fresh.p-t\endcsname{0.999}
\expandafter\def\csname cnt@rev.pol.perm.n.sspro.p2.eqcost-lin-vs-uniform-fresh.p-t\endcsname{1}
\expandafter\def\csname pct@rev.pol.perm.n.sspro.p2.eqcost-lin-vs-uniform-fresh.p-t\endcsname{99.9}
\expandafter\def\csname ppt@rev.pol.perm.n.sspro.p2.eqcost-lin-vs-uniform-fresh.p-t\endcsname{$+$99.9}
\expandafter\def\csname val@rev.pol.perm.n.sspro.p2.eqcost-lin-vs-uniform-fresh.p-t-lt\endcsname{0.001}
\expandafter\def\csname cnt@rev.pol.perm.n.sspro.p2.eqcost-lin-vs-uniform-fresh.p-t-lt\endcsname{0}
\expandafter\def\csname pct@rev.pol.perm.n.sspro.p2.eqcost-lin-vs-uniform-fresh.p-t-lt\endcsname{0.1}
\expandafter\def\csname ppt@rev.pol.perm.n.sspro.p2.eqcost-lin-vs-uniform-fresh.p-t-lt\endcsname{$+$0.1}
\expandafter\def\csname val@rev.pol.perm.n.sspro.p2.rcr-vs-full-recert-ni.p-signflip\endcsname{0.911}
\expandafter\def\csname cnt@rev.pol.perm.n.sspro.p2.rcr-vs-full-recert-ni.p-signflip\endcsname{1}
\expandafter\def\csname pct@rev.pol.perm.n.sspro.p2.rcr-vs-full-recert-ni.p-signflip\endcsname{91.1}
\expandafter\def\csname ppt@rev.pol.perm.n.sspro.p2.rcr-vs-full-recert-ni.p-signflip\endcsname{$+$91.1}
\expandafter\def\csname val@rev.pol.perm.n.sspro.p2.rcr-vs-full-recert-ni.p-t\endcsname{0.913}
\expandafter\def\csname cnt@rev.pol.perm.n.sspro.p2.rcr-vs-full-recert-ni.p-t\endcsname{1}
\expandafter\def\csname pct@rev.pol.perm.n.sspro.p2.rcr-vs-full-recert-ni.p-t\endcsname{91.3}
\expandafter\def\csname ppt@rev.pol.perm.n.sspro.p2.rcr-vs-full-recert-ni.p-t\endcsname{$+$91.3}
\expandafter\def\csname val@rev.pol.perm.n.sspro.p2.rcr-vs-generic-ppi-source.p-signflip\endcsname{0.564}
\expandafter\def\csname cnt@rev.pol.perm.n.sspro.p2.rcr-vs-generic-ppi-source.p-signflip\endcsname{1}
\expandafter\def\csname pct@rev.pol.perm.n.sspro.p2.rcr-vs-generic-ppi-source.p-signflip\endcsname{56.4}
\expandafter\def\csname ppt@rev.pol.perm.n.sspro.p2.rcr-vs-generic-ppi-source.p-signflip\endcsname{$+$56.4}
\expandafter\def\csname val@rev.pol.perm.n.sspro.p2.rcr-vs-generic-ppi-source.p-signflip-holm\endcsname{1.000}
\expandafter\def\csname cnt@rev.pol.perm.n.sspro.p2.rcr-vs-generic-ppi-source.p-signflip-holm\endcsname{1}
\expandafter\def\csname pct@rev.pol.perm.n.sspro.p2.rcr-vs-generic-ppi-source.p-signflip-holm\endcsname{100.0}
\expandafter\def\csname ppt@rev.pol.perm.n.sspro.p2.rcr-vs-generic-ppi-source.p-signflip-holm\endcsname{$+$100.0}
\expandafter\def\csname val@rev.pol.perm.n.sspro.p2.rcr-vs-generic-ppi-source.p-t\endcsname{0.566}
\expandafter\def\csname cnt@rev.pol.perm.n.sspro.p2.rcr-vs-generic-ppi-source.p-t\endcsname{1}
\expandafter\def\csname pct@rev.pol.perm.n.sspro.p2.rcr-vs-generic-ppi-source.p-t\endcsname{56.6}
\expandafter\def\csname ppt@rev.pol.perm.n.sspro.p2.rcr-vs-generic-ppi-source.p-t\endcsname{$+$56.6}
\expandafter\def\csname val@rev.pol.perm.n.sspro.p2.rcr-vs-generic-ppi-source.p-t-holm\endcsname{0.864}
\expandafter\def\csname cnt@rev.pol.perm.n.sspro.p2.rcr-vs-generic-ppi-source.p-t-holm\endcsname{1}
\expandafter\def\csname pct@rev.pol.perm.n.sspro.p2.rcr-vs-generic-ppi-source.p-t-holm\endcsname{86.4}
\expandafter\def\csname ppt@rev.pol.perm.n.sspro.p2.rcr-vs-generic-ppi-source.p-t-holm\endcsname{$+$86.4}
\expandafter\def\csname val@rev.pol.perm.n.sspro.p2.rcr-vs-replay-ppi-uniform.p-signflip\endcsname{0.500}
\expandafter\def\csname cnt@rev.pol.perm.n.sspro.p2.rcr-vs-replay-ppi-uniform.p-signflip\endcsname{0}
\expandafter\def\csname pct@rev.pol.perm.n.sspro.p2.rcr-vs-replay-ppi-uniform.p-signflip\endcsname{50.0}
\expandafter\def\csname ppt@rev.pol.perm.n.sspro.p2.rcr-vs-replay-ppi-uniform.p-signflip\endcsname{$+$50.0}
\expandafter\def\csname val@rev.pol.perm.n.sspro.p2.rcr-vs-replay-ppi-uniform.p-signflip-holm\endcsname{1.000}
\expandafter\def\csname cnt@rev.pol.perm.n.sspro.p2.rcr-vs-replay-ppi-uniform.p-signflip-holm\endcsname{1}
\expandafter\def\csname pct@rev.pol.perm.n.sspro.p2.rcr-vs-replay-ppi-uniform.p-signflip-holm\endcsname{100.0}
\expandafter\def\csname ppt@rev.pol.perm.n.sspro.p2.rcr-vs-replay-ppi-uniform.p-signflip-holm\endcsname{$+$100.0}
\expandafter\def\csname val@rev.pol.perm.n.sspro.p2.rcr-vs-replay-ppi-uniform.p-t\endcsname{0.432}
\expandafter\def\csname cnt@rev.pol.perm.n.sspro.p2.rcr-vs-replay-ppi-uniform.p-t\endcsname{0}
\expandafter\def\csname pct@rev.pol.perm.n.sspro.p2.rcr-vs-replay-ppi-uniform.p-t\endcsname{43.2}
\expandafter\def\csname ppt@rev.pol.perm.n.sspro.p2.rcr-vs-replay-ppi-uniform.p-t\endcsname{$+$43.2}
\expandafter\def\csname val@rev.pol.perm.n.sspro.p2.rcr-vs-replay-ppi-uniform.p-t-holm\endcsname{0.864}
\expandafter\def\csname cnt@rev.pol.perm.n.sspro.p2.rcr-vs-replay-ppi-uniform.p-t-holm\endcsname{1}
\expandafter\def\csname pct@rev.pol.perm.n.sspro.p2.rcr-vs-replay-ppi-uniform.p-t-holm\endcsname{86.4}
\expandafter\def\csname ppt@rev.pol.perm.n.sspro.p2.rcr-vs-replay-ppi-uniform.p-t-holm\endcsname{$+$86.4}
\expandafter\def\csname val@rev.pol.perm.n.ttest-gt.n-rejected\endcsname{2}
\expandafter\def\csname cnt@rev.pol.perm.n.ttest-gt.n-rejected\endcsname{2}
\expandafter\def\csname pct@rev.pol.perm.n.ttest-gt.n-rejected\endcsname{200.0}
\expandafter\def\csname ppt@rev.pol.perm.n.ttest-gt.n-rejected\endcsname{$+$200.0}
\expandafter\def\csname val@rev.pol.perm.n.uiv.p1.eqcost-lin-vs-best-single.p-signflip\endcsname{0.878}
\expandafter\def\csname cnt@rev.pol.perm.n.uiv.p1.eqcost-lin-vs-best-single.p-signflip\endcsname{1}
\expandafter\def\csname pct@rev.pol.perm.n.uiv.p1.eqcost-lin-vs-best-single.p-signflip\endcsname{87.8}
\expandafter\def\csname ppt@rev.pol.perm.n.uiv.p1.eqcost-lin-vs-best-single.p-signflip\endcsname{$+$87.8}
\expandafter\def\csname val@rev.pol.perm.n.uiv.p1.eqcost-lin-vs-best-single.p-signflip-lt\endcsname{0.122}
\expandafter\def\csname cnt@rev.pol.perm.n.uiv.p1.eqcost-lin-vs-best-single.p-signflip-lt\endcsname{0}
\expandafter\def\csname pct@rev.pol.perm.n.uiv.p1.eqcost-lin-vs-best-single.p-signflip-lt\endcsname{12.2}
\expandafter\def\csname ppt@rev.pol.perm.n.uiv.p1.eqcost-lin-vs-best-single.p-signflip-lt\endcsname{$+$12.2}
\expandafter\def\csname val@rev.pol.perm.n.uiv.p1.eqcost-lin-vs-best-single.p-t\endcsname{0.883}
\expandafter\def\csname cnt@rev.pol.perm.n.uiv.p1.eqcost-lin-vs-best-single.p-t\endcsname{1}
\expandafter\def\csname pct@rev.pol.perm.n.uiv.p1.eqcost-lin-vs-best-single.p-t\endcsname{88.3}
\expandafter\def\csname ppt@rev.pol.perm.n.uiv.p1.eqcost-lin-vs-best-single.p-t\endcsname{$+$88.3}
\expandafter\def\csname val@rev.pol.perm.n.uiv.p1.eqcost-lin-vs-best-single.p-t-lt\endcsname{0.117}
\expandafter\def\csname cnt@rev.pol.perm.n.uiv.p1.eqcost-lin-vs-best-single.p-t-lt\endcsname{0}
\expandafter\def\csname pct@rev.pol.perm.n.uiv.p1.eqcost-lin-vs-best-single.p-t-lt\endcsname{11.7}
\expandafter\def\csname ppt@rev.pol.perm.n.uiv.p1.eqcost-lin-vs-best-single.p-t-lt\endcsname{$+$11.7}
\expandafter\def\csname val@rev.pol.perm.n.uiv.p1.eqcost-lin-vs-generic-ppi-source.p-signflip\endcsname{0.000}
\expandafter\def\csname cnt@rev.pol.perm.n.uiv.p1.eqcost-lin-vs-generic-ppi-source.p-signflip\endcsname{0}
\expandafter\def\csname pct@rev.pol.perm.n.uiv.p1.eqcost-lin-vs-generic-ppi-source.p-signflip\endcsname{0.0}
\expandafter\def\csname ppt@rev.pol.perm.n.uiv.p1.eqcost-lin-vs-generic-ppi-source.p-signflip\endcsname{0.0}
\expandafter\def\csname val@rev.pol.perm.n.uiv.p1.eqcost-lin-vs-generic-ppi-source.p-signflip-lt\endcsname{1.000}
\expandafter\def\csname cnt@rev.pol.perm.n.uiv.p1.eqcost-lin-vs-generic-ppi-source.p-signflip-lt\endcsname{1}
\expandafter\def\csname pct@rev.pol.perm.n.uiv.p1.eqcost-lin-vs-generic-ppi-source.p-signflip-lt\endcsname{100.0}
\expandafter\def\csname ppt@rev.pol.perm.n.uiv.p1.eqcost-lin-vs-generic-ppi-source.p-signflip-lt\endcsname{$+$100.0}
\expandafter\def\csname val@rev.pol.perm.n.uiv.p1.eqcost-lin-vs-generic-ppi-source.p-t\endcsname{0.000}
\expandafter\def\csname cnt@rev.pol.perm.n.uiv.p1.eqcost-lin-vs-generic-ppi-source.p-t\endcsname{0}
\expandafter\def\csname pct@rev.pol.perm.n.uiv.p1.eqcost-lin-vs-generic-ppi-source.p-t\endcsname{0.0}
\expandafter\def\csname ppt@rev.pol.perm.n.uiv.p1.eqcost-lin-vs-generic-ppi-source.p-t\endcsname{0.0}
\expandafter\def\csname val@rev.pol.perm.n.uiv.p1.eqcost-lin-vs-generic-ppi-source.p-t-lt\endcsname{1.000}
\expandafter\def\csname cnt@rev.pol.perm.n.uiv.p1.eqcost-lin-vs-generic-ppi-source.p-t-lt\endcsname{1}
\expandafter\def\csname pct@rev.pol.perm.n.uiv.p1.eqcost-lin-vs-generic-ppi-source.p-t-lt\endcsname{100.0}
\expandafter\def\csname ppt@rev.pol.perm.n.uiv.p1.eqcost-lin-vs-generic-ppi-source.p-t-lt\endcsname{$+$100.0}
\expandafter\def\csname val@rev.pol.perm.n.uiv.p1.eqcost-lin-vs-replay-ppi-uniform.p-signflip\endcsname{0.001}
\expandafter\def\csname cnt@rev.pol.perm.n.uiv.p1.eqcost-lin-vs-replay-ppi-uniform.p-signflip\endcsname{0}
\expandafter\def\csname pct@rev.pol.perm.n.uiv.p1.eqcost-lin-vs-replay-ppi-uniform.p-signflip\endcsname{0.1}
\expandafter\def\csname ppt@rev.pol.perm.n.uiv.p1.eqcost-lin-vs-replay-ppi-uniform.p-signflip\endcsname{$+$0.1}
\expandafter\def\csname val@rev.pol.perm.n.uiv.p1.eqcost-lin-vs-replay-ppi-uniform.p-signflip-lt\endcsname{0.999}
\expandafter\def\csname cnt@rev.pol.perm.n.uiv.p1.eqcost-lin-vs-replay-ppi-uniform.p-signflip-lt\endcsname{1}
\expandafter\def\csname pct@rev.pol.perm.n.uiv.p1.eqcost-lin-vs-replay-ppi-uniform.p-signflip-lt\endcsname{99.9}
\expandafter\def\csname ppt@rev.pol.perm.n.uiv.p1.eqcost-lin-vs-replay-ppi-uniform.p-signflip-lt\endcsname{$+$99.9}
\expandafter\def\csname val@rev.pol.perm.n.uiv.p1.eqcost-lin-vs-replay-ppi-uniform.p-t\endcsname{0.001}
\expandafter\def\csname cnt@rev.pol.perm.n.uiv.p1.eqcost-lin-vs-replay-ppi-uniform.p-t\endcsname{0}
\expandafter\def\csname pct@rev.pol.perm.n.uiv.p1.eqcost-lin-vs-replay-ppi-uniform.p-t\endcsname{0.1}
\expandafter\def\csname ppt@rev.pol.perm.n.uiv.p1.eqcost-lin-vs-replay-ppi-uniform.p-t\endcsname{$+$0.1}
\expandafter\def\csname val@rev.pol.perm.n.uiv.p1.eqcost-lin-vs-replay-ppi-uniform.p-t-lt\endcsname{0.999}
\expandafter\def\csname cnt@rev.pol.perm.n.uiv.p1.eqcost-lin-vs-replay-ppi-uniform.p-t-lt\endcsname{1}
\expandafter\def\csname pct@rev.pol.perm.n.uiv.p1.eqcost-lin-vs-replay-ppi-uniform.p-t-lt\endcsname{99.9}
\expandafter\def\csname ppt@rev.pol.perm.n.uiv.p1.eqcost-lin-vs-replay-ppi-uniform.p-t-lt\endcsname{$+$99.9}
\expandafter\def\csname val@rev.pol.perm.n.uiv.p1.eqcost-lin-vs-uniform-fresh.p-signflip\endcsname{0.000}
\expandafter\def\csname cnt@rev.pol.perm.n.uiv.p1.eqcost-lin-vs-uniform-fresh.p-signflip\endcsname{0}
\expandafter\def\csname pct@rev.pol.perm.n.uiv.p1.eqcost-lin-vs-uniform-fresh.p-signflip\endcsname{0.0}
\expandafter\def\csname ppt@rev.pol.perm.n.uiv.p1.eqcost-lin-vs-uniform-fresh.p-signflip\endcsname{0.0}
\expandafter\def\csname val@rev.pol.perm.n.uiv.p1.eqcost-lin-vs-uniform-fresh.p-signflip-lt\endcsname{1.000}
\expandafter\def\csname cnt@rev.pol.perm.n.uiv.p1.eqcost-lin-vs-uniform-fresh.p-signflip-lt\endcsname{1}
\expandafter\def\csname pct@rev.pol.perm.n.uiv.p1.eqcost-lin-vs-uniform-fresh.p-signflip-lt\endcsname{100.0}
\expandafter\def\csname ppt@rev.pol.perm.n.uiv.p1.eqcost-lin-vs-uniform-fresh.p-signflip-lt\endcsname{$+$100.0}
\expandafter\def\csname val@rev.pol.perm.n.uiv.p1.eqcost-lin-vs-uniform-fresh.p-t\endcsname{0.001}
\expandafter\def\csname cnt@rev.pol.perm.n.uiv.p1.eqcost-lin-vs-uniform-fresh.p-t\endcsname{0}
\expandafter\def\csname pct@rev.pol.perm.n.uiv.p1.eqcost-lin-vs-uniform-fresh.p-t\endcsname{0.1}
\expandafter\def\csname ppt@rev.pol.perm.n.uiv.p1.eqcost-lin-vs-uniform-fresh.p-t\endcsname{$+$0.1}
\expandafter\def\csname val@rev.pol.perm.n.uiv.p1.eqcost-lin-vs-uniform-fresh.p-t-lt\endcsname{0.999}
\expandafter\def\csname cnt@rev.pol.perm.n.uiv.p1.eqcost-lin-vs-uniform-fresh.p-t-lt\endcsname{1}
\expandafter\def\csname pct@rev.pol.perm.n.uiv.p1.eqcost-lin-vs-uniform-fresh.p-t-lt\endcsname{99.9}
\expandafter\def\csname ppt@rev.pol.perm.n.uiv.p1.eqcost-lin-vs-uniform-fresh.p-t-lt\endcsname{$+$99.9}
\expandafter\def\csname val@rev.pol.perm.n.uiv.p1.rcr-vs-full-recert-ni.p-signflip\endcsname{0.000}
\expandafter\def\csname cnt@rev.pol.perm.n.uiv.p1.rcr-vs-full-recert-ni.p-signflip\endcsname{0}
\expandafter\def\csname pct@rev.pol.perm.n.uiv.p1.rcr-vs-full-recert-ni.p-signflip\endcsname{0.0}
\expandafter\def\csname ppt@rev.pol.perm.n.uiv.p1.rcr-vs-full-recert-ni.p-signflip\endcsname{0.0}
\expandafter\def\csname val@rev.pol.perm.n.uiv.p1.rcr-vs-full-recert-ni.p-t\endcsname{0.000}
\expandafter\def\csname cnt@rev.pol.perm.n.uiv.p1.rcr-vs-full-recert-ni.p-t\endcsname{0}
\expandafter\def\csname pct@rev.pol.perm.n.uiv.p1.rcr-vs-full-recert-ni.p-t\endcsname{0.0}
\expandafter\def\csname ppt@rev.pol.perm.n.uiv.p1.rcr-vs-full-recert-ni.p-t\endcsname{0.0}
\expandafter\def\csname val@rev.pol.perm.n.uiv.p1.rcr-vs-generic-ppi-source.p-signflip\endcsname{0.000}
\expandafter\def\csname cnt@rev.pol.perm.n.uiv.p1.rcr-vs-generic-ppi-source.p-signflip\endcsname{0}
\expandafter\def\csname pct@rev.pol.perm.n.uiv.p1.rcr-vs-generic-ppi-source.p-signflip\endcsname{0.0}
\expandafter\def\csname ppt@rev.pol.perm.n.uiv.p1.rcr-vs-generic-ppi-source.p-signflip\endcsname{0.0}
\expandafter\def\csname val@rev.pol.perm.n.uiv.p1.rcr-vs-generic-ppi-source.p-signflip-holm\endcsname{0.004}
\expandafter\def\csname cnt@rev.pol.perm.n.uiv.p1.rcr-vs-generic-ppi-source.p-signflip-holm\endcsname{0}
\expandafter\def\csname pct@rev.pol.perm.n.uiv.p1.rcr-vs-generic-ppi-source.p-signflip-holm\endcsname{0.4}
\expandafter\def\csname ppt@rev.pol.perm.n.uiv.p1.rcr-vs-generic-ppi-source.p-signflip-holm\endcsname{$+$0.4}
\expandafter\def\csname val@rev.pol.perm.n.uiv.p1.rcr-vs-generic-ppi-source.p-t\endcsname{0.001}
\expandafter\def\csname cnt@rev.pol.perm.n.uiv.p1.rcr-vs-generic-ppi-source.p-t\endcsname{0}
\expandafter\def\csname pct@rev.pol.perm.n.uiv.p1.rcr-vs-generic-ppi-source.p-t\endcsname{0.1}
\expandafter\def\csname ppt@rev.pol.perm.n.uiv.p1.rcr-vs-generic-ppi-source.p-t\endcsname{$+$0.1}
\expandafter\def\csname val@rev.pol.perm.n.uiv.p1.rcr-vs-generic-ppi-source.p-t-holm\endcsname{0.008}
\expandafter\def\csname cnt@rev.pol.perm.n.uiv.p1.rcr-vs-generic-ppi-source.p-t-holm\endcsname{0}
\expandafter\def\csname pct@rev.pol.perm.n.uiv.p1.rcr-vs-generic-ppi-source.p-t-holm\endcsname{0.8}
\expandafter\def\csname ppt@rev.pol.perm.n.uiv.p1.rcr-vs-generic-ppi-source.p-t-holm\endcsname{$+$0.8}
\expandafter\def\csname val@rev.pol.perm.n.uiv.p1.rcr-vs-replay-ppi-uniform.p-signflip\endcsname{0.001}
\expandafter\def\csname cnt@rev.pol.perm.n.uiv.p1.rcr-vs-replay-ppi-uniform.p-signflip\endcsname{0}
\expandafter\def\csname pct@rev.pol.perm.n.uiv.p1.rcr-vs-replay-ppi-uniform.p-signflip\endcsname{0.1}
\expandafter\def\csname ppt@rev.pol.perm.n.uiv.p1.rcr-vs-replay-ppi-uniform.p-signflip\endcsname{$+$0.1}
\expandafter\def\csname val@rev.pol.perm.n.uiv.p1.rcr-vs-replay-ppi-uniform.p-signflip-holm\endcsname{0.005}
\expandafter\def\csname cnt@rev.pol.perm.n.uiv.p1.rcr-vs-replay-ppi-uniform.p-signflip-holm\endcsname{0}
\expandafter\def\csname pct@rev.pol.perm.n.uiv.p1.rcr-vs-replay-ppi-uniform.p-signflip-holm\endcsname{0.5}
\expandafter\def\csname ppt@rev.pol.perm.n.uiv.p1.rcr-vs-replay-ppi-uniform.p-signflip-holm\endcsname{$+$0.5}
\expandafter\def\csname val@rev.pol.perm.n.uiv.p1.rcr-vs-replay-ppi-uniform.p-t\endcsname{0.001}
\expandafter\def\csname cnt@rev.pol.perm.n.uiv.p1.rcr-vs-replay-ppi-uniform.p-t\endcsname{0}
\expandafter\def\csname pct@rev.pol.perm.n.uiv.p1.rcr-vs-replay-ppi-uniform.p-t\endcsname{0.1}
\expandafter\def\csname ppt@rev.pol.perm.n.uiv.p1.rcr-vs-replay-ppi-uniform.p-t\endcsname{$+$0.1}
\expandafter\def\csname val@rev.pol.perm.n.uiv.p1.rcr-vs-replay-ppi-uniform.p-t-holm\endcsname{0.007}
\expandafter\def\csname cnt@rev.pol.perm.n.uiv.p1.rcr-vs-replay-ppi-uniform.p-t-holm\endcsname{0}
\expandafter\def\csname pct@rev.pol.perm.n.uiv.p1.rcr-vs-replay-ppi-uniform.p-t-holm\endcsname{0.7}
\expandafter\def\csname ppt@rev.pol.perm.n.uiv.p1.rcr-vs-replay-ppi-uniform.p-t-holm\endcsname{$+$0.7}
\expandafter\def\csname val@rev.pol.perm.n.uiv.p2.eqcost-lin-vs-best-single.p-signflip\endcsname{0.227}
\expandafter\def\csname cnt@rev.pol.perm.n.uiv.p2.eqcost-lin-vs-best-single.p-signflip\endcsname{0}
\expandafter\def\csname pct@rev.pol.perm.n.uiv.p2.eqcost-lin-vs-best-single.p-signflip\endcsname{22.7}
\expandafter\def\csname ppt@rev.pol.perm.n.uiv.p2.eqcost-lin-vs-best-single.p-signflip\endcsname{$+$22.7}
\expandafter\def\csname val@rev.pol.perm.n.uiv.p2.eqcost-lin-vs-best-single.p-signflip-lt\endcsname{0.774}
\expandafter\def\csname cnt@rev.pol.perm.n.uiv.p2.eqcost-lin-vs-best-single.p-signflip-lt\endcsname{1}
\expandafter\def\csname pct@rev.pol.perm.n.uiv.p2.eqcost-lin-vs-best-single.p-signflip-lt\endcsname{77.4}
\expandafter\def\csname ppt@rev.pol.perm.n.uiv.p2.eqcost-lin-vs-best-single.p-signflip-lt\endcsname{$+$77.4}
\expandafter\def\csname val@rev.pol.perm.n.uiv.p2.eqcost-lin-vs-best-single.p-t\endcsname{0.216}
\expandafter\def\csname cnt@rev.pol.perm.n.uiv.p2.eqcost-lin-vs-best-single.p-t\endcsname{0}
\expandafter\def\csname pct@rev.pol.perm.n.uiv.p2.eqcost-lin-vs-best-single.p-t\endcsname{21.6}
\expandafter\def\csname ppt@rev.pol.perm.n.uiv.p2.eqcost-lin-vs-best-single.p-t\endcsname{$+$21.6}
\expandafter\def\csname val@rev.pol.perm.n.uiv.p2.eqcost-lin-vs-best-single.p-t-lt\endcsname{0.784}
\expandafter\def\csname cnt@rev.pol.perm.n.uiv.p2.eqcost-lin-vs-best-single.p-t-lt\endcsname{1}
\expandafter\def\csname pct@rev.pol.perm.n.uiv.p2.eqcost-lin-vs-best-single.p-t-lt\endcsname{78.4}
\expandafter\def\csname ppt@rev.pol.perm.n.uiv.p2.eqcost-lin-vs-best-single.p-t-lt\endcsname{$+$78.4}
\expandafter\def\csname val@rev.pol.perm.n.uiv.p2.eqcost-lin-vs-generic-ppi-source.p-signflip\endcsname{0.513}
\expandafter\def\csname cnt@rev.pol.perm.n.uiv.p2.eqcost-lin-vs-generic-ppi-source.p-signflip\endcsname{1}
\expandafter\def\csname pct@rev.pol.perm.n.uiv.p2.eqcost-lin-vs-generic-ppi-source.p-signflip\endcsname{51.3}
\expandafter\def\csname ppt@rev.pol.perm.n.uiv.p2.eqcost-lin-vs-generic-ppi-source.p-signflip\endcsname{$+$51.3}
\expandafter\def\csname val@rev.pol.perm.n.uiv.p2.eqcost-lin-vs-generic-ppi-source.p-signflip-lt\endcsname{0.488}
\expandafter\def\csname cnt@rev.pol.perm.n.uiv.p2.eqcost-lin-vs-generic-ppi-source.p-signflip-lt\endcsname{0}
\expandafter\def\csname pct@rev.pol.perm.n.uiv.p2.eqcost-lin-vs-generic-ppi-source.p-signflip-lt\endcsname{48.8}
\expandafter\def\csname ppt@rev.pol.perm.n.uiv.p2.eqcost-lin-vs-generic-ppi-source.p-signflip-lt\endcsname{$+$48.8}
\expandafter\def\csname val@rev.pol.perm.n.uiv.p2.eqcost-lin-vs-generic-ppi-source.p-t\endcsname{0.516}
\expandafter\def\csname cnt@rev.pol.perm.n.uiv.p2.eqcost-lin-vs-generic-ppi-source.p-t\endcsname{1}
\expandafter\def\csname pct@rev.pol.perm.n.uiv.p2.eqcost-lin-vs-generic-ppi-source.p-t\endcsname{51.6}
\expandafter\def\csname ppt@rev.pol.perm.n.uiv.p2.eqcost-lin-vs-generic-ppi-source.p-t\endcsname{$+$51.6}
\expandafter\def\csname val@rev.pol.perm.n.uiv.p2.eqcost-lin-vs-generic-ppi-source.p-t-lt\endcsname{0.484}
\expandafter\def\csname cnt@rev.pol.perm.n.uiv.p2.eqcost-lin-vs-generic-ppi-source.p-t-lt\endcsname{0}
\expandafter\def\csname pct@rev.pol.perm.n.uiv.p2.eqcost-lin-vs-generic-ppi-source.p-t-lt\endcsname{48.4}
\expandafter\def\csname ppt@rev.pol.perm.n.uiv.p2.eqcost-lin-vs-generic-ppi-source.p-t-lt\endcsname{$+$48.4}
\expandafter\def\csname val@rev.pol.perm.n.uiv.p2.eqcost-lin-vs-replay-ppi-uniform.p-signflip\endcsname{0.009}
\expandafter\def\csname cnt@rev.pol.perm.n.uiv.p2.eqcost-lin-vs-replay-ppi-uniform.p-signflip\endcsname{0}
\expandafter\def\csname pct@rev.pol.perm.n.uiv.p2.eqcost-lin-vs-replay-ppi-uniform.p-signflip\endcsname{0.9}
\expandafter\def\csname ppt@rev.pol.perm.n.uiv.p2.eqcost-lin-vs-replay-ppi-uniform.p-signflip\endcsname{$+$0.9}
\expandafter\def\csname val@rev.pol.perm.n.uiv.p2.eqcost-lin-vs-replay-ppi-uniform.p-signflip-lt\endcsname{0.991}
\expandafter\def\csname cnt@rev.pol.perm.n.uiv.p2.eqcost-lin-vs-replay-ppi-uniform.p-signflip-lt\endcsname{1}
\expandafter\def\csname pct@rev.pol.perm.n.uiv.p2.eqcost-lin-vs-replay-ppi-uniform.p-signflip-lt\endcsname{99.1}
\expandafter\def\csname ppt@rev.pol.perm.n.uiv.p2.eqcost-lin-vs-replay-ppi-uniform.p-signflip-lt\endcsname{$+$99.1}
\expandafter\def\csname val@rev.pol.perm.n.uiv.p2.eqcost-lin-vs-replay-ppi-uniform.p-t\endcsname{0.010}
\expandafter\def\csname cnt@rev.pol.perm.n.uiv.p2.eqcost-lin-vs-replay-ppi-uniform.p-t\endcsname{0}
\expandafter\def\csname pct@rev.pol.perm.n.uiv.p2.eqcost-lin-vs-replay-ppi-uniform.p-t\endcsname{1.0}
\expandafter\def\csname ppt@rev.pol.perm.n.uiv.p2.eqcost-lin-vs-replay-ppi-uniform.p-t\endcsname{$+$1.0}
\expandafter\def\csname val@rev.pol.perm.n.uiv.p2.eqcost-lin-vs-replay-ppi-uniform.p-t-lt\endcsname{0.990}
\expandafter\def\csname cnt@rev.pol.perm.n.uiv.p2.eqcost-lin-vs-replay-ppi-uniform.p-t-lt\endcsname{1}
\expandafter\def\csname pct@rev.pol.perm.n.uiv.p2.eqcost-lin-vs-replay-ppi-uniform.p-t-lt\endcsname{99.0}
\expandafter\def\csname ppt@rev.pol.perm.n.uiv.p2.eqcost-lin-vs-replay-ppi-uniform.p-t-lt\endcsname{$+$99.0}
\expandafter\def\csname val@rev.pol.perm.n.uiv.p2.eqcost-lin-vs-uniform-fresh.p-signflip\endcsname{0.211}
\expandafter\def\csname cnt@rev.pol.perm.n.uiv.p2.eqcost-lin-vs-uniform-fresh.p-signflip\endcsname{0}
\expandafter\def\csname pct@rev.pol.perm.n.uiv.p2.eqcost-lin-vs-uniform-fresh.p-signflip\endcsname{21.1}
\expandafter\def\csname ppt@rev.pol.perm.n.uiv.p2.eqcost-lin-vs-uniform-fresh.p-signflip\endcsname{$+$21.1}
\expandafter\def\csname val@rev.pol.perm.n.uiv.p2.eqcost-lin-vs-uniform-fresh.p-signflip-lt\endcsname{0.790}
\expandafter\def\csname cnt@rev.pol.perm.n.uiv.p2.eqcost-lin-vs-uniform-fresh.p-signflip-lt\endcsname{1}
\expandafter\def\csname pct@rev.pol.perm.n.uiv.p2.eqcost-lin-vs-uniform-fresh.p-signflip-lt\endcsname{79.0}
\expandafter\def\csname ppt@rev.pol.perm.n.uiv.p2.eqcost-lin-vs-uniform-fresh.p-signflip-lt\endcsname{$+$79.0}
\expandafter\def\csname val@rev.pol.perm.n.uiv.p2.eqcost-lin-vs-uniform-fresh.p-t\endcsname{0.202}
\expandafter\def\csname cnt@rev.pol.perm.n.uiv.p2.eqcost-lin-vs-uniform-fresh.p-t\endcsname{0}
\expandafter\def\csname pct@rev.pol.perm.n.uiv.p2.eqcost-lin-vs-uniform-fresh.p-t\endcsname{20.2}
\expandafter\def\csname ppt@rev.pol.perm.n.uiv.p2.eqcost-lin-vs-uniform-fresh.p-t\endcsname{$+$20.2}
\expandafter\def\csname val@rev.pol.perm.n.uiv.p2.eqcost-lin-vs-uniform-fresh.p-t-lt\endcsname{0.798}
\expandafter\def\csname cnt@rev.pol.perm.n.uiv.p2.eqcost-lin-vs-uniform-fresh.p-t-lt\endcsname{1}
\expandafter\def\csname pct@rev.pol.perm.n.uiv.p2.eqcost-lin-vs-uniform-fresh.p-t-lt\endcsname{79.8}
\expandafter\def\csname ppt@rev.pol.perm.n.uiv.p2.eqcost-lin-vs-uniform-fresh.p-t-lt\endcsname{$+$79.8}
\expandafter\def\csname val@rev.pol.perm.n.uiv.p2.rcr-vs-full-recert-ni.p-signflip\endcsname{0.010}
\expandafter\def\csname cnt@rev.pol.perm.n.uiv.p2.rcr-vs-full-recert-ni.p-signflip\endcsname{0}
\expandafter\def\csname pct@rev.pol.perm.n.uiv.p2.rcr-vs-full-recert-ni.p-signflip\endcsname{1.0}
\expandafter\def\csname ppt@rev.pol.perm.n.uiv.p2.rcr-vs-full-recert-ni.p-signflip\endcsname{$+$1.0}
\expandafter\def\csname val@rev.pol.perm.n.uiv.p2.rcr-vs-full-recert-ni.p-t\endcsname{0.013}
\expandafter\def\csname cnt@rev.pol.perm.n.uiv.p2.rcr-vs-full-recert-ni.p-t\endcsname{0}
\expandafter\def\csname pct@rev.pol.perm.n.uiv.p2.rcr-vs-full-recert-ni.p-t\endcsname{1.3}
\expandafter\def\csname ppt@rev.pol.perm.n.uiv.p2.rcr-vs-full-recert-ni.p-t\endcsname{$+$1.3}
\expandafter\def\csname val@rev.pol.perm.n.uiv.p2.rcr-vs-generic-ppi-source.p-signflip\endcsname{0.056}
\expandafter\def\csname cnt@rev.pol.perm.n.uiv.p2.rcr-vs-generic-ppi-source.p-signflip\endcsname{0}
\expandafter\def\csname pct@rev.pol.perm.n.uiv.p2.rcr-vs-generic-ppi-source.p-signflip\endcsname{5.6}
\expandafter\def\csname ppt@rev.pol.perm.n.uiv.p2.rcr-vs-generic-ppi-source.p-signflip\endcsname{$+$5.6}
\expandafter\def\csname val@rev.pol.perm.n.uiv.p2.rcr-vs-generic-ppi-source.p-signflip-holm\endcsname{0.168}
\expandafter\def\csname cnt@rev.pol.perm.n.uiv.p2.rcr-vs-generic-ppi-source.p-signflip-holm\endcsname{0}
\expandafter\def\csname pct@rev.pol.perm.n.uiv.p2.rcr-vs-generic-ppi-source.p-signflip-holm\endcsname{16.8}
\expandafter\def\csname ppt@rev.pol.perm.n.uiv.p2.rcr-vs-generic-ppi-source.p-signflip-holm\endcsname{$+$16.8}
\expandafter\def\csname val@rev.pol.perm.n.uiv.p2.rcr-vs-generic-ppi-source.p-t\endcsname{0.057}
\expandafter\def\csname cnt@rev.pol.perm.n.uiv.p2.rcr-vs-generic-ppi-source.p-t\endcsname{0}
\expandafter\def\csname pct@rev.pol.perm.n.uiv.p2.rcr-vs-generic-ppi-source.p-t\endcsname{5.7}
\expandafter\def\csname ppt@rev.pol.perm.n.uiv.p2.rcr-vs-generic-ppi-source.p-t\endcsname{$+$5.7}
\expandafter\def\csname val@rev.pol.perm.n.uiv.p2.rcr-vs-generic-ppi-source.p-t-holm\endcsname{0.214}
\expandafter\def\csname cnt@rev.pol.perm.n.uiv.p2.rcr-vs-generic-ppi-source.p-t-holm\endcsname{0}
\expandafter\def\csname pct@rev.pol.perm.n.uiv.p2.rcr-vs-generic-ppi-source.p-t-holm\endcsname{21.4}
\expandafter\def\csname ppt@rev.pol.perm.n.uiv.p2.rcr-vs-generic-ppi-source.p-t-holm\endcsname{$+$21.4}
\expandafter\def\csname val@rev.pol.perm.n.uiv.p2.rcr-vs-replay-ppi-uniform.p-signflip\endcsname{0.010}
\expandafter\def\csname cnt@rev.pol.perm.n.uiv.p2.rcr-vs-replay-ppi-uniform.p-signflip\endcsname{0}
\expandafter\def\csname pct@rev.pol.perm.n.uiv.p2.rcr-vs-replay-ppi-uniform.p-signflip\endcsname{1.0}
\expandafter\def\csname ppt@rev.pol.perm.n.uiv.p2.rcr-vs-replay-ppi-uniform.p-signflip\endcsname{$+$1.0}
\expandafter\def\csname val@rev.pol.perm.n.uiv.p2.rcr-vs-replay-ppi-uniform.p-signflip-holm\endcsname{0.061}
\expandafter\def\csname cnt@rev.pol.perm.n.uiv.p2.rcr-vs-replay-ppi-uniform.p-signflip-holm\endcsname{0}
\expandafter\def\csname pct@rev.pol.perm.n.uiv.p2.rcr-vs-replay-ppi-uniform.p-signflip-holm\endcsname{6.1}
\expandafter\def\csname ppt@rev.pol.perm.n.uiv.p2.rcr-vs-replay-ppi-uniform.p-signflip-holm\endcsname{$+$6.1}
\expandafter\def\csname val@rev.pol.perm.n.uiv.p2.rcr-vs-replay-ppi-uniform.p-t\endcsname{0.011}
\expandafter\def\csname cnt@rev.pol.perm.n.uiv.p2.rcr-vs-replay-ppi-uniform.p-t\endcsname{0}
\expandafter\def\csname pct@rev.pol.perm.n.uiv.p2.rcr-vs-replay-ppi-uniform.p-t\endcsname{1.1}
\expandafter\def\csname ppt@rev.pol.perm.n.uiv.p2.rcr-vs-replay-ppi-uniform.p-t\endcsname{$+$1.1}
\expandafter\def\csname val@rev.pol.perm.n.uiv.p2.rcr-vs-replay-ppi-uniform.p-t-holm\endcsname{0.063}
\expandafter\def\csname cnt@rev.pol.perm.n.uiv.p2.rcr-vs-replay-ppi-uniform.p-t-holm\endcsname{0}
\expandafter\def\csname pct@rev.pol.perm.n.uiv.p2.rcr-vs-replay-ppi-uniform.p-t-holm\endcsname{6.3}
\expandafter\def\csname ppt@rev.pol.perm.n.uiv.p2.rcr-vs-replay-ppi-uniform.p-t-holm\endcsname{$+$6.3}
\expandafter\def\csname val@rev.pol.rules.gain.lam0.5.max\endcsname{0.026}
\expandafter\def\csname cnt@rev.pol.rules.gain.lam0.5.max\endcsname{0}
\expandafter\def\csname pct@rev.pol.rules.gain.lam0.5.max\endcsname{2.6}
\expandafter\def\csname ppt@rev.pol.rules.gain.lam0.5.max\endcsname{$+$2.6}
\expandafter\def\csname val@rev.pol.rules.gain.lam0.5.min\endcsname{-0.032}
\expandafter\def\csname cnt@rev.pol.rules.gain.lam0.5.min\endcsname{0}
\expandafter\def\csname pct@rev.pol.rules.gain.lam0.5.min\endcsname{$-$3.2}
\expandafter\def\csname ppt@rev.pol.rules.gain.lam0.5.min\endcsname{$-$3.2}
\expandafter\def\csname val@rev.pol.rules.gain.lam0.5.n\endcsname{28}
\expandafter\def\csname cnt@rev.pol.rules.gain.lam0.5.n\endcsname{28}
\expandafter\def\csname pct@rev.pol.rules.gain.lam0.5.n\endcsname{2800.0}
\expandafter\def\csname ppt@rev.pol.rules.gain.lam0.5.n\endcsname{$+$2800.0}
\expandafter\def\csname val@rev.pol.rules.gain.lam0.5.n-negative\endcsname{4}
\expandafter\def\csname cnt@rev.pol.rules.gain.lam0.5.n-negative\endcsname{4}
\expandafter\def\csname pct@rev.pol.rules.gain.lam0.5.n-negative\endcsname{400.0}
\expandafter\def\csname ppt@rev.pol.rules.gain.lam0.5.n-negative\endcsname{$+$400.0}
\expandafter\def\csname val@rev.pol.rules.gain.lam0.5.n-positive\endcsname{9}
\expandafter\def\csname cnt@rev.pol.rules.gain.lam0.5.n-positive\endcsname{9}
\expandafter\def\csname pct@rev.pol.rules.gain.lam0.5.n-positive\endcsname{900.0}
\expandafter\def\csname ppt@rev.pol.rules.gain.lam0.5.n-positive\endcsname{$+$900.0}
\expandafter\def\csname val@rev.pol.rules.gain.lam1.max\endcsname{0.026}
\expandafter\def\csname cnt@rev.pol.rules.gain.lam1.max\endcsname{0}
\expandafter\def\csname pct@rev.pol.rules.gain.lam1.max\endcsname{2.6}
\expandafter\def\csname ppt@rev.pol.rules.gain.lam1.max\endcsname{$+$2.6}
\expandafter\def\csname val@rev.pol.rules.gain.lam1.min\endcsname{-0.032}
\expandafter\def\csname cnt@rev.pol.rules.gain.lam1.min\endcsname{0}
\expandafter\def\csname pct@rev.pol.rules.gain.lam1.min\endcsname{$-$3.2}
\expandafter\def\csname ppt@rev.pol.rules.gain.lam1.min\endcsname{$-$3.2}
\expandafter\def\csname val@rev.pol.rules.gain.lam1.n\endcsname{28}
\expandafter\def\csname cnt@rev.pol.rules.gain.lam1.n\endcsname{28}
\expandafter\def\csname pct@rev.pol.rules.gain.lam1.n\endcsname{2800.0}
\expandafter\def\csname ppt@rev.pol.rules.gain.lam1.n\endcsname{$+$2800.0}
\expandafter\def\csname val@rev.pol.rules.gain.lam1.n-negative\endcsname{4}
\expandafter\def\csname cnt@rev.pol.rules.gain.lam1.n-negative\endcsname{4}
\expandafter\def\csname pct@rev.pol.rules.gain.lam1.n-negative\endcsname{400.0}
\expandafter\def\csname ppt@rev.pol.rules.gain.lam1.n-negative\endcsname{$+$400.0}
\expandafter\def\csname val@rev.pol.rules.gain.lam1.n-positive\endcsname{11}
\expandafter\def\csname cnt@rev.pol.rules.gain.lam1.n-positive\endcsname{11}
\expandafter\def\csname pct@rev.pol.rules.gain.lam1.n-positive\endcsname{1100.0}
\expandafter\def\csname ppt@rev.pol.rules.gain.lam1.n-positive\endcsname{$+$1100.0}
\expandafter\def\csname val@rev.pol.rules.gain.m0.5.max\endcsname{0.026}
\expandafter\def\csname cnt@rev.pol.rules.gain.m0.5.max\endcsname{0}
\expandafter\def\csname pct@rev.pol.rules.gain.m0.5.max\endcsname{2.6}
\expandafter\def\csname ppt@rev.pol.rules.gain.m0.5.max\endcsname{$+$2.6}
\expandafter\def\csname val@rev.pol.rules.gain.m0.5.min\endcsname{-0.032}
\expandafter\def\csname cnt@rev.pol.rules.gain.m0.5.min\endcsname{0}
\expandafter\def\csname pct@rev.pol.rules.gain.m0.5.min\endcsname{$-$3.2}
\expandafter\def\csname ppt@rev.pol.rules.gain.m0.5.min\endcsname{$-$3.2}
\expandafter\def\csname val@rev.pol.rules.gain.m0.5.n\endcsname{28}
\expandafter\def\csname cnt@rev.pol.rules.gain.m0.5.n\endcsname{28}
\expandafter\def\csname pct@rev.pol.rules.gain.m0.5.n\endcsname{2800.0}
\expandafter\def\csname ppt@rev.pol.rules.gain.m0.5.n\endcsname{$+$2800.0}
\expandafter\def\csname val@rev.pol.rules.gain.m0.5.n-negative\endcsname{3}
\expandafter\def\csname cnt@rev.pol.rules.gain.m0.5.n-negative\endcsname{3}
\expandafter\def\csname pct@rev.pol.rules.gain.m0.5.n-negative\endcsname{300.0}
\expandafter\def\csname ppt@rev.pol.rules.gain.m0.5.n-negative\endcsname{$+$300.0}
\expandafter\def\csname val@rev.pol.rules.gain.m0.5.n-positive\endcsname{7}
\expandafter\def\csname cnt@rev.pol.rules.gain.m0.5.n-positive\endcsname{7}
\expandafter\def\csname pct@rev.pol.rules.gain.m0.5.n-positive\endcsname{700.0}
\expandafter\def\csname ppt@rev.pol.rules.gain.m0.5.n-positive\endcsname{$+$700.0}
\expandafter\def\csname val@rev.pol.rules.gain.m1.max\endcsname{0.026}
\expandafter\def\csname cnt@rev.pol.rules.gain.m1.max\endcsname{0}
\expandafter\def\csname pct@rev.pol.rules.gain.m1.max\endcsname{2.6}
\expandafter\def\csname ppt@rev.pol.rules.gain.m1.max\endcsname{$+$2.6}
\expandafter\def\csname val@rev.pol.rules.gain.m1.min\endcsname{-0.032}
\expandafter\def\csname cnt@rev.pol.rules.gain.m1.min\endcsname{0}
\expandafter\def\csname pct@rev.pol.rules.gain.m1.min\endcsname{$-$3.2}
\expandafter\def\csname ppt@rev.pol.rules.gain.m1.min\endcsname{$-$3.2}
\expandafter\def\csname val@rev.pol.rules.gain.m1.n\endcsname{28}
\expandafter\def\csname cnt@rev.pol.rules.gain.m1.n\endcsname{28}
\expandafter\def\csname pct@rev.pol.rules.gain.m1.n\endcsname{2800.0}
\expandafter\def\csname ppt@rev.pol.rules.gain.m1.n\endcsname{$+$2800.0}
\expandafter\def\csname val@rev.pol.rules.gain.m1.n-negative\endcsname{4}
\expandafter\def\csname cnt@rev.pol.rules.gain.m1.n-negative\endcsname{4}
\expandafter\def\csname pct@rev.pol.rules.gain.m1.n-negative\endcsname{400.0}
\expandafter\def\csname ppt@rev.pol.rules.gain.m1.n-negative\endcsname{$+$400.0}
\expandafter\def\csname val@rev.pol.rules.gain.m1.n-positive\endcsname{7}
\expandafter\def\csname cnt@rev.pol.rules.gain.m1.n-positive\endcsname{7}
\expandafter\def\csname pct@rev.pol.rules.gain.m1.n-positive\endcsname{700.0}
\expandafter\def\csname ppt@rev.pol.rules.gain.m1.n-positive\endcsname{$+$700.0}
\expandafter\def\csname val@rev.pol.rules.gain.m2.max\endcsname{0.040}
\expandafter\def\csname cnt@rev.pol.rules.gain.m2.max\endcsname{0}
\expandafter\def\csname pct@rev.pol.rules.gain.m2.max\endcsname{4.0}
\expandafter\def\csname ppt@rev.pol.rules.gain.m2.max\endcsname{$+$4.0}
\expandafter\def\csname val@rev.pol.rules.gain.m2.min\endcsname{-0.032}
\expandafter\def\csname cnt@rev.pol.rules.gain.m2.min\endcsname{0}
\expandafter\def\csname pct@rev.pol.rules.gain.m2.min\endcsname{$-$3.2}
\expandafter\def\csname ppt@rev.pol.rules.gain.m2.min\endcsname{$-$3.2}
\expandafter\def\csname val@rev.pol.rules.gain.m2.n\endcsname{28}
\expandafter\def\csname cnt@rev.pol.rules.gain.m2.n\endcsname{28}
\expandafter\def\csname pct@rev.pol.rules.gain.m2.n\endcsname{2800.0}
\expandafter\def\csname ppt@rev.pol.rules.gain.m2.n\endcsname{$+$2800.0}
\expandafter\def\csname val@rev.pol.rules.gain.m2.n-negative\endcsname{4}
\expandafter\def\csname cnt@rev.pol.rules.gain.m2.n-negative\endcsname{4}
\expandafter\def\csname pct@rev.pol.rules.gain.m2.n-negative\endcsname{400.0}
\expandafter\def\csname ppt@rev.pol.rules.gain.m2.n-negative\endcsname{$+$400.0}
\expandafter\def\csname val@rev.pol.rules.gain.m2.n-positive\endcsname{12}
\expandafter\def\csname cnt@rev.pol.rules.gain.m2.n-positive\endcsname{12}
\expandafter\def\csname pct@rev.pol.rules.gain.m2.n-positive\endcsname{1200.0}
\expandafter\def\csname ppt@rev.pol.rules.gain.m2.n-positive\endcsname{$+$1200.0}
\expandafter\def\csname val@rev.pol.rules.gain.sspro.lam0.5.max\endcsname{0.000}
\expandafter\def\csname cnt@rev.pol.rules.gain.sspro.lam0.5.max\endcsname{0}
\expandafter\def\csname pct@rev.pol.rules.gain.sspro.lam0.5.max\endcsname{0.0}
\expandafter\def\csname ppt@rev.pol.rules.gain.sspro.lam0.5.max\endcsname{0.0}
\expandafter\def\csname val@rev.pol.rules.gain.sspro.lam0.5.min\endcsname{-0.032}
\expandafter\def\csname cnt@rev.pol.rules.gain.sspro.lam0.5.min\endcsname{0}
\expandafter\def\csname pct@rev.pol.rules.gain.sspro.lam0.5.min\endcsname{$-$3.2}
\expandafter\def\csname ppt@rev.pol.rules.gain.sspro.lam0.5.min\endcsname{$-$3.2}
\expandafter\def\csname val@rev.pol.rules.gain.sspro.lam1.max\endcsname{0.000}
\expandafter\def\csname cnt@rev.pol.rules.gain.sspro.lam1.max\endcsname{0}
\expandafter\def\csname pct@rev.pol.rules.gain.sspro.lam1.max\endcsname{0.0}
\expandafter\def\csname ppt@rev.pol.rules.gain.sspro.lam1.max\endcsname{0.0}
\expandafter\def\csname val@rev.pol.rules.gain.sspro.lam1.min\endcsname{-0.032}
\expandafter\def\csname cnt@rev.pol.rules.gain.sspro.lam1.min\endcsname{0}
\expandafter\def\csname pct@rev.pol.rules.gain.sspro.lam1.min\endcsname{$-$3.2}
\expandafter\def\csname ppt@rev.pol.rules.gain.sspro.lam1.min\endcsname{$-$3.2}
\expandafter\def\csname val@rev.pol.rules.gain.sspro.m0.5.max\endcsname{0.000}
\expandafter\def\csname cnt@rev.pol.rules.gain.sspro.m0.5.max\endcsname{0}
\expandafter\def\csname pct@rev.pol.rules.gain.sspro.m0.5.max\endcsname{0.0}
\expandafter\def\csname ppt@rev.pol.rules.gain.sspro.m0.5.max\endcsname{0.0}
\expandafter\def\csname val@rev.pol.rules.gain.sspro.m0.5.min\endcsname{-0.032}
\expandafter\def\csname cnt@rev.pol.rules.gain.sspro.m0.5.min\endcsname{0}
\expandafter\def\csname pct@rev.pol.rules.gain.sspro.m0.5.min\endcsname{$-$3.2}
\expandafter\def\csname ppt@rev.pol.rules.gain.sspro.m0.5.min\endcsname{$-$3.2}
\expandafter\def\csname val@rev.pol.rules.gain.sspro.m1.max\endcsname{0.000}
\expandafter\def\csname cnt@rev.pol.rules.gain.sspro.m1.max\endcsname{0}
\expandafter\def\csname pct@rev.pol.rules.gain.sspro.m1.max\endcsname{0.0}
\expandafter\def\csname ppt@rev.pol.rules.gain.sspro.m1.max\endcsname{0.0}
\expandafter\def\csname val@rev.pol.rules.gain.sspro.m1.min\endcsname{-0.032}
\expandafter\def\csname cnt@rev.pol.rules.gain.sspro.m1.min\endcsname{0}
\expandafter\def\csname pct@rev.pol.rules.gain.sspro.m1.min\endcsname{$-$3.2}
\expandafter\def\csname ppt@rev.pol.rules.gain.sspro.m1.min\endcsname{$-$3.2}
\expandafter\def\csname val@rev.pol.rules.gain.sspro.m2.max\endcsname{0.000}
\expandafter\def\csname cnt@rev.pol.rules.gain.sspro.m2.max\endcsname{0}
\expandafter\def\csname pct@rev.pol.rules.gain.sspro.m2.max\endcsname{0.0}
\expandafter\def\csname ppt@rev.pol.rules.gain.sspro.m2.max\endcsname{0.0}
\expandafter\def\csname val@rev.pol.rules.gain.sspro.m2.min\endcsname{-0.032}
\expandafter\def\csname cnt@rev.pol.rules.gain.sspro.m2.min\endcsname{0}
\expandafter\def\csname pct@rev.pol.rules.gain.sspro.m2.min\endcsname{$-$3.2}
\expandafter\def\csname ppt@rev.pol.rules.gain.sspro.m2.min\endcsname{$-$3.2}
\expandafter\def\csname val@rev.pol.rules.gain.uiv.lam0.5.max\endcsname{0.026}
\expandafter\def\csname cnt@rev.pol.rules.gain.uiv.lam0.5.max\endcsname{0}
\expandafter\def\csname pct@rev.pol.rules.gain.uiv.lam0.5.max\endcsname{2.6}
\expandafter\def\csname ppt@rev.pol.rules.gain.uiv.lam0.5.max\endcsname{$+$2.6}
\expandafter\def\csname val@rev.pol.rules.gain.uiv.lam0.5.min\endcsname{0.000}
\expandafter\def\csname cnt@rev.pol.rules.gain.uiv.lam0.5.min\endcsname{0}
\expandafter\def\csname pct@rev.pol.rules.gain.uiv.lam0.5.min\endcsname{0.0}
\expandafter\def\csname ppt@rev.pol.rules.gain.uiv.lam0.5.min\endcsname{0.0}
\expandafter\def\csname val@rev.pol.rules.gain.uiv.lam1.max\endcsname{0.026}
\expandafter\def\csname cnt@rev.pol.rules.gain.uiv.lam1.max\endcsname{0}
\expandafter\def\csname pct@rev.pol.rules.gain.uiv.lam1.max\endcsname{2.6}
\expandafter\def\csname ppt@rev.pol.rules.gain.uiv.lam1.max\endcsname{$+$2.6}
\expandafter\def\csname val@rev.pol.rules.gain.uiv.lam1.min\endcsname{0.000}
\expandafter\def\csname cnt@rev.pol.rules.gain.uiv.lam1.min\endcsname{0}
\expandafter\def\csname pct@rev.pol.rules.gain.uiv.lam1.min\endcsname{0.0}
\expandafter\def\csname ppt@rev.pol.rules.gain.uiv.lam1.min\endcsname{0.0}
\expandafter\def\csname val@rev.pol.rules.gain.uiv.m0.5.max\endcsname{0.026}
\expandafter\def\csname cnt@rev.pol.rules.gain.uiv.m0.5.max\endcsname{0}
\expandafter\def\csname pct@rev.pol.rules.gain.uiv.m0.5.max\endcsname{2.6}
\expandafter\def\csname ppt@rev.pol.rules.gain.uiv.m0.5.max\endcsname{$+$2.6}
\expandafter\def\csname val@rev.pol.rules.gain.uiv.m0.5.min\endcsname{0.000}
\expandafter\def\csname cnt@rev.pol.rules.gain.uiv.m0.5.min\endcsname{0}
\expandafter\def\csname pct@rev.pol.rules.gain.uiv.m0.5.min\endcsname{0.0}
\expandafter\def\csname ppt@rev.pol.rules.gain.uiv.m0.5.min\endcsname{0.0}
\expandafter\def\csname val@rev.pol.rules.gain.uiv.m1.max\endcsname{0.026}
\expandafter\def\csname cnt@rev.pol.rules.gain.uiv.m1.max\endcsname{0}
\expandafter\def\csname pct@rev.pol.rules.gain.uiv.m1.max\endcsname{2.6}
\expandafter\def\csname ppt@rev.pol.rules.gain.uiv.m1.max\endcsname{$+$2.6}
\expandafter\def\csname val@rev.pol.rules.gain.uiv.m1.min\endcsname{0.000}
\expandafter\def\csname cnt@rev.pol.rules.gain.uiv.m1.min\endcsname{0}
\expandafter\def\csname pct@rev.pol.rules.gain.uiv.m1.min\endcsname{0.0}
\expandafter\def\csname ppt@rev.pol.rules.gain.uiv.m1.min\endcsname{0.0}
\expandafter\def\csname val@rev.pol.rules.gain.uiv.m2.max\endcsname{0.040}
\expandafter\def\csname cnt@rev.pol.rules.gain.uiv.m2.max\endcsname{0}
\expandafter\def\csname pct@rev.pol.rules.gain.uiv.m2.max\endcsname{4.0}
\expandafter\def\csname ppt@rev.pol.rules.gain.uiv.m2.max\endcsname{$+$4.0}
\expandafter\def\csname val@rev.pol.rules.gain.uiv.m2.min\endcsname{0.000}
\expandafter\def\csname cnt@rev.pol.rules.gain.uiv.m2.min\endcsname{0}
\expandafter\def\csname pct@rev.pol.rules.gain.uiv.m2.min\endcsname{0.0}
\expandafter\def\csname ppt@rev.pol.rules.gain.uiv.m2.min\endcsname{0.0}
\expandafter\def\csname val@rev.pol.rules.sspro.p1.full-recert.frozen.keep\endcsname{1.000}
\expandafter\def\csname cnt@rev.pol.rules.sspro.p1.full-recert.frozen.keep\endcsname{1}
\expandafter\def\csname pct@rev.pol.rules.sspro.p1.full-recert.frozen.keep\endcsname{100.0}
\expandafter\def\csname ppt@rev.pol.rules.sspro.p1.full-recert.frozen.keep\endcsname{$+$100.0}
\expandafter\def\csname val@rev.pol.rules.sspro.p1.full-recert.frozen.req\endcsname{2.716}
\expandafter\def\csname cnt@rev.pol.rules.sspro.p1.full-recert.frozen.req\endcsname{3}
\expandafter\def\csname pct@rev.pol.rules.sspro.p1.full-recert.frozen.req\endcsname{271.6}
\expandafter\def\csname ppt@rev.pol.rules.sspro.p1.full-recert.frozen.req\endcsname{$+$271.6}
\expandafter\def\csname val@rev.pol.rules.sspro.p1.full-recert.frozen.score\endcsname{0.859}
\expandafter\def\csname cnt@rev.pol.rules.sspro.p1.full-recert.frozen.score\endcsname{1}
\expandafter\def\csname pct@rev.pol.rules.sspro.p1.full-recert.frozen.score\endcsname{85.9}
\expandafter\def\csname ppt@rev.pol.rules.sspro.p1.full-recert.frozen.score\endcsname{$+$85.9}
\expandafter\def\csname val@rev.pol.rules.sspro.p1.full-recert.lam0.5.keep\endcsname{1.000}
\expandafter\def\csname cnt@rev.pol.rules.sspro.p1.full-recert.lam0.5.keep\endcsname{1}
\expandafter\def\csname pct@rev.pol.rules.sspro.p1.full-recert.lam0.5.keep\endcsname{100.0}
\expandafter\def\csname ppt@rev.pol.rules.sspro.p1.full-recert.lam0.5.keep\endcsname{$+$100.0}
\expandafter\def\csname val@rev.pol.rules.sspro.p1.full-recert.lam0.5.req\endcsname{2.716}
\expandafter\def\csname cnt@rev.pol.rules.sspro.p1.full-recert.lam0.5.req\endcsname{3}
\expandafter\def\csname pct@rev.pol.rules.sspro.p1.full-recert.lam0.5.req\endcsname{271.6}
\expandafter\def\csname ppt@rev.pol.rules.sspro.p1.full-recert.lam0.5.req\endcsname{$+$271.6}
\expandafter\def\csname val@rev.pol.rules.sspro.p1.full-recert.lam0.5.score\endcsname{0.859}
\expandafter\def\csname cnt@rev.pol.rules.sspro.p1.full-recert.lam0.5.score\endcsname{1}
\expandafter\def\csname pct@rev.pol.rules.sspro.p1.full-recert.lam0.5.score\endcsname{85.9}
\expandafter\def\csname ppt@rev.pol.rules.sspro.p1.full-recert.lam0.5.score\endcsname{$+$85.9}
\expandafter\def\csname val@rev.pol.rules.sspro.p1.full-recert.lam1.keep\endcsname{1.000}
\expandafter\def\csname cnt@rev.pol.rules.sspro.p1.full-recert.lam1.keep\endcsname{1}
\expandafter\def\csname pct@rev.pol.rules.sspro.p1.full-recert.lam1.keep\endcsname{100.0}
\expandafter\def\csname ppt@rev.pol.rules.sspro.p1.full-recert.lam1.keep\endcsname{$+$100.0}
\expandafter\def\csname val@rev.pol.rules.sspro.p1.full-recert.lam1.req\endcsname{2.716}
\expandafter\def\csname cnt@rev.pol.rules.sspro.p1.full-recert.lam1.req\endcsname{3}
\expandafter\def\csname pct@rev.pol.rules.sspro.p1.full-recert.lam1.req\endcsname{271.6}
\expandafter\def\csname ppt@rev.pol.rules.sspro.p1.full-recert.lam1.req\endcsname{$+$271.6}
\expandafter\def\csname val@rev.pol.rules.sspro.p1.full-recert.lam1.score\endcsname{0.859}
\expandafter\def\csname cnt@rev.pol.rules.sspro.p1.full-recert.lam1.score\endcsname{1}
\expandafter\def\csname pct@rev.pol.rules.sspro.p1.full-recert.lam1.score\endcsname{85.9}
\expandafter\def\csname ppt@rev.pol.rules.sspro.p1.full-recert.lam1.score\endcsname{$+$85.9}
\expandafter\def\csname val@rev.pol.rules.sspro.p1.full-recert.m0.5.keep\endcsname{1.000}
\expandafter\def\csname cnt@rev.pol.rules.sspro.p1.full-recert.m0.5.keep\endcsname{1}
\expandafter\def\csname pct@rev.pol.rules.sspro.p1.full-recert.m0.5.keep\endcsname{100.0}
\expandafter\def\csname ppt@rev.pol.rules.sspro.p1.full-recert.m0.5.keep\endcsname{$+$100.0}
\expandafter\def\csname val@rev.pol.rules.sspro.p1.full-recert.m0.5.req\endcsname{2.716}
\expandafter\def\csname cnt@rev.pol.rules.sspro.p1.full-recert.m0.5.req\endcsname{3}
\expandafter\def\csname pct@rev.pol.rules.sspro.p1.full-recert.m0.5.req\endcsname{271.6}
\expandafter\def\csname ppt@rev.pol.rules.sspro.p1.full-recert.m0.5.req\endcsname{$+$271.6}
\expandafter\def\csname val@rev.pol.rules.sspro.p1.full-recert.m0.5.score\endcsname{0.859}
\expandafter\def\csname cnt@rev.pol.rules.sspro.p1.full-recert.m0.5.score\endcsname{1}
\expandafter\def\csname pct@rev.pol.rules.sspro.p1.full-recert.m0.5.score\endcsname{85.9}
\expandafter\def\csname ppt@rev.pol.rules.sspro.p1.full-recert.m0.5.score\endcsname{$+$85.9}
\expandafter\def\csname val@rev.pol.rules.sspro.p1.full-recert.m1.keep\endcsname{1.000}
\expandafter\def\csname cnt@rev.pol.rules.sspro.p1.full-recert.m1.keep\endcsname{1}
\expandafter\def\csname pct@rev.pol.rules.sspro.p1.full-recert.m1.keep\endcsname{100.0}
\expandafter\def\csname ppt@rev.pol.rules.sspro.p1.full-recert.m1.keep\endcsname{$+$100.0}
\expandafter\def\csname val@rev.pol.rules.sspro.p1.full-recert.m1.req\endcsname{2.716}
\expandafter\def\csname cnt@rev.pol.rules.sspro.p1.full-recert.m1.req\endcsname{3}
\expandafter\def\csname pct@rev.pol.rules.sspro.p1.full-recert.m1.req\endcsname{271.6}
\expandafter\def\csname ppt@rev.pol.rules.sspro.p1.full-recert.m1.req\endcsname{$+$271.6}
\expandafter\def\csname val@rev.pol.rules.sspro.p1.full-recert.m1.score\endcsname{0.859}
\expandafter\def\csname cnt@rev.pol.rules.sspro.p1.full-recert.m1.score\endcsname{1}
\expandafter\def\csname pct@rev.pol.rules.sspro.p1.full-recert.m1.score\endcsname{85.9}
\expandafter\def\csname ppt@rev.pol.rules.sspro.p1.full-recert.m1.score\endcsname{$+$85.9}
\expandafter\def\csname val@rev.pol.rules.sspro.p1.full-recert.m2.keep\endcsname{1.000}
\expandafter\def\csname cnt@rev.pol.rules.sspro.p1.full-recert.m2.keep\endcsname{1}
\expandafter\def\csname pct@rev.pol.rules.sspro.p1.full-recert.m2.keep\endcsname{100.0}
\expandafter\def\csname ppt@rev.pol.rules.sspro.p1.full-recert.m2.keep\endcsname{$+$100.0}
\expandafter\def\csname val@rev.pol.rules.sspro.p1.full-recert.m2.req\endcsname{2.716}
\expandafter\def\csname cnt@rev.pol.rules.sspro.p1.full-recert.m2.req\endcsname{3}
\expandafter\def\csname pct@rev.pol.rules.sspro.p1.full-recert.m2.req\endcsname{271.6}
\expandafter\def\csname ppt@rev.pol.rules.sspro.p1.full-recert.m2.req\endcsname{$+$271.6}
\expandafter\def\csname val@rev.pol.rules.sspro.p1.full-recert.m2.score\endcsname{0.859}
\expandafter\def\csname cnt@rev.pol.rules.sspro.p1.full-recert.m2.score\endcsname{1}
\expandafter\def\csname pct@rev.pol.rules.sspro.p1.full-recert.m2.score\endcsname{85.9}
\expandafter\def\csname ppt@rev.pol.rules.sspro.p1.full-recert.m2.score\endcsname{$+$85.9}
\expandafter\def\csname val@rev.pol.rules.sspro.p1.generic-ppi-source@0.25.frozen.keep\endcsname{0.901}
\expandafter\def\csname cnt@rev.pol.rules.sspro.p1.generic-ppi-source@0.25.frozen.keep\endcsname{1}
\expandafter\def\csname pct@rev.pol.rules.sspro.p1.generic-ppi-source@0.25.frozen.keep\endcsname{90.1}
\expandafter\def\csname ppt@rev.pol.rules.sspro.p1.generic-ppi-source@0.25.frozen.keep\endcsname{$+$90.1}
\expandafter\def\csname val@rev.pol.rules.sspro.p1.generic-ppi-source@0.25.frozen.req\endcsname{2.579}
\expandafter\def\csname cnt@rev.pol.rules.sspro.p1.generic-ppi-source@0.25.frozen.req\endcsname{3}
\expandafter\def\csname pct@rev.pol.rules.sspro.p1.generic-ppi-source@0.25.frozen.req\endcsname{257.9}
\expandafter\def\csname ppt@rev.pol.rules.sspro.p1.generic-ppi-source@0.25.frozen.req\endcsname{$+$257.9}
\expandafter\def\csname val@rev.pol.rules.sspro.p1.generic-ppi-source@0.25.frozen.score\endcsname{0.847}
\expandafter\def\csname cnt@rev.pol.rules.sspro.p1.generic-ppi-source@0.25.frozen.score\endcsname{1}
\expandafter\def\csname pct@rev.pol.rules.sspro.p1.generic-ppi-source@0.25.frozen.score\endcsname{84.7}
\expandafter\def\csname ppt@rev.pol.rules.sspro.p1.generic-ppi-source@0.25.frozen.score\endcsname{$+$84.7}
\expandafter\def\csname val@rev.pol.rules.sspro.p1.generic-ppi-source@0.25.lam0.5.keep\endcsname{0.901}
\expandafter\def\csname cnt@rev.pol.rules.sspro.p1.generic-ppi-source@0.25.lam0.5.keep\endcsname{1}
\expandafter\def\csname pct@rev.pol.rules.sspro.p1.generic-ppi-source@0.25.lam0.5.keep\endcsname{90.1}
\expandafter\def\csname ppt@rev.pol.rules.sspro.p1.generic-ppi-source@0.25.lam0.5.keep\endcsname{$+$90.1}
\expandafter\def\csname val@rev.pol.rules.sspro.p1.generic-ppi-source@0.25.lam0.5.req\endcsname{2.579}
\expandafter\def\csname cnt@rev.pol.rules.sspro.p1.generic-ppi-source@0.25.lam0.5.req\endcsname{3}
\expandafter\def\csname pct@rev.pol.rules.sspro.p1.generic-ppi-source@0.25.lam0.5.req\endcsname{257.9}
\expandafter\def\csname ppt@rev.pol.rules.sspro.p1.generic-ppi-source@0.25.lam0.5.req\endcsname{$+$257.9}
\expandafter\def\csname val@rev.pol.rules.sspro.p1.generic-ppi-source@0.25.lam0.5.score\endcsname{0.847}
\expandafter\def\csname cnt@rev.pol.rules.sspro.p1.generic-ppi-source@0.25.lam0.5.score\endcsname{1}
\expandafter\def\csname pct@rev.pol.rules.sspro.p1.generic-ppi-source@0.25.lam0.5.score\endcsname{84.7}
\expandafter\def\csname ppt@rev.pol.rules.sspro.p1.generic-ppi-source@0.25.lam0.5.score\endcsname{$+$84.7}
\expandafter\def\csname val@rev.pol.rules.sspro.p1.generic-ppi-source@0.25.lam1.keep\endcsname{0.901}
\expandafter\def\csname cnt@rev.pol.rules.sspro.p1.generic-ppi-source@0.25.lam1.keep\endcsname{1}
\expandafter\def\csname pct@rev.pol.rules.sspro.p1.generic-ppi-source@0.25.lam1.keep\endcsname{90.1}
\expandafter\def\csname ppt@rev.pol.rules.sspro.p1.generic-ppi-source@0.25.lam1.keep\endcsname{$+$90.1}
\expandafter\def\csname val@rev.pol.rules.sspro.p1.generic-ppi-source@0.25.lam1.req\endcsname{2.579}
\expandafter\def\csname cnt@rev.pol.rules.sspro.p1.generic-ppi-source@0.25.lam1.req\endcsname{3}
\expandafter\def\csname pct@rev.pol.rules.sspro.p1.generic-ppi-source@0.25.lam1.req\endcsname{257.9}
\expandafter\def\csname ppt@rev.pol.rules.sspro.p1.generic-ppi-source@0.25.lam1.req\endcsname{$+$257.9}
\expandafter\def\csname val@rev.pol.rules.sspro.p1.generic-ppi-source@0.25.lam1.score\endcsname{0.847}
\expandafter\def\csname cnt@rev.pol.rules.sspro.p1.generic-ppi-source@0.25.lam1.score\endcsname{1}
\expandafter\def\csname pct@rev.pol.rules.sspro.p1.generic-ppi-source@0.25.lam1.score\endcsname{84.7}
\expandafter\def\csname ppt@rev.pol.rules.sspro.p1.generic-ppi-source@0.25.lam1.score\endcsname{$+$84.7}
\expandafter\def\csname val@rev.pol.rules.sspro.p1.generic-ppi-source@0.25.m0.5.keep\endcsname{0.901}
\expandafter\def\csname cnt@rev.pol.rules.sspro.p1.generic-ppi-source@0.25.m0.5.keep\endcsname{1}
\expandafter\def\csname pct@rev.pol.rules.sspro.p1.generic-ppi-source@0.25.m0.5.keep\endcsname{90.1}
\expandafter\def\csname ppt@rev.pol.rules.sspro.p1.generic-ppi-source@0.25.m0.5.keep\endcsname{$+$90.1}
\expandafter\def\csname val@rev.pol.rules.sspro.p1.generic-ppi-source@0.25.m0.5.req\endcsname{2.579}
\expandafter\def\csname cnt@rev.pol.rules.sspro.p1.generic-ppi-source@0.25.m0.5.req\endcsname{3}
\expandafter\def\csname pct@rev.pol.rules.sspro.p1.generic-ppi-source@0.25.m0.5.req\endcsname{257.9}
\expandafter\def\csname ppt@rev.pol.rules.sspro.p1.generic-ppi-source@0.25.m0.5.req\endcsname{$+$257.9}
\expandafter\def\csname val@rev.pol.rules.sspro.p1.generic-ppi-source@0.25.m0.5.score\endcsname{0.847}
\expandafter\def\csname cnt@rev.pol.rules.sspro.p1.generic-ppi-source@0.25.m0.5.score\endcsname{1}
\expandafter\def\csname pct@rev.pol.rules.sspro.p1.generic-ppi-source@0.25.m0.5.score\endcsname{84.7}
\expandafter\def\csname ppt@rev.pol.rules.sspro.p1.generic-ppi-source@0.25.m0.5.score\endcsname{$+$84.7}
\expandafter\def\csname val@rev.pol.rules.sspro.p1.generic-ppi-source@0.25.m1.keep\endcsname{0.901}
\expandafter\def\csname cnt@rev.pol.rules.sspro.p1.generic-ppi-source@0.25.m1.keep\endcsname{1}
\expandafter\def\csname pct@rev.pol.rules.sspro.p1.generic-ppi-source@0.25.m1.keep\endcsname{90.1}
\expandafter\def\csname ppt@rev.pol.rules.sspro.p1.generic-ppi-source@0.25.m1.keep\endcsname{$+$90.1}
\expandafter\def\csname val@rev.pol.rules.sspro.p1.generic-ppi-source@0.25.m1.req\endcsname{2.579}
\expandafter\def\csname cnt@rev.pol.rules.sspro.p1.generic-ppi-source@0.25.m1.req\endcsname{3}
\expandafter\def\csname pct@rev.pol.rules.sspro.p1.generic-ppi-source@0.25.m1.req\endcsname{257.9}
\expandafter\def\csname ppt@rev.pol.rules.sspro.p1.generic-ppi-source@0.25.m1.req\endcsname{$+$257.9}
\expandafter\def\csname val@rev.pol.rules.sspro.p1.generic-ppi-source@0.25.m1.score\endcsname{0.847}
\expandafter\def\csname cnt@rev.pol.rules.sspro.p1.generic-ppi-source@0.25.m1.score\endcsname{1}
\expandafter\def\csname pct@rev.pol.rules.sspro.p1.generic-ppi-source@0.25.m1.score\endcsname{84.7}
\expandafter\def\csname ppt@rev.pol.rules.sspro.p1.generic-ppi-source@0.25.m1.score\endcsname{$+$84.7}
\expandafter\def\csname val@rev.pol.rules.sspro.p1.generic-ppi-source@0.25.m2.keep\endcsname{0.901}
\expandafter\def\csname cnt@rev.pol.rules.sspro.p1.generic-ppi-source@0.25.m2.keep\endcsname{1}
\expandafter\def\csname pct@rev.pol.rules.sspro.p1.generic-ppi-source@0.25.m2.keep\endcsname{90.1}
\expandafter\def\csname ppt@rev.pol.rules.sspro.p1.generic-ppi-source@0.25.m2.keep\endcsname{$+$90.1}
\expandafter\def\csname val@rev.pol.rules.sspro.p1.generic-ppi-source@0.25.m2.req\endcsname{2.579}
\expandafter\def\csname cnt@rev.pol.rules.sspro.p1.generic-ppi-source@0.25.m2.req\endcsname{3}
\expandafter\def\csname pct@rev.pol.rules.sspro.p1.generic-ppi-source@0.25.m2.req\endcsname{257.9}
\expandafter\def\csname ppt@rev.pol.rules.sspro.p1.generic-ppi-source@0.25.m2.req\endcsname{$+$257.9}
\expandafter\def\csname val@rev.pol.rules.sspro.p1.generic-ppi-source@0.25.m2.score\endcsname{0.847}
\expandafter\def\csname cnt@rev.pol.rules.sspro.p1.generic-ppi-source@0.25.m2.score\endcsname{1}
\expandafter\def\csname pct@rev.pol.rules.sspro.p1.generic-ppi-source@0.25.m2.score\endcsname{84.7}
\expandafter\def\csname ppt@rev.pol.rules.sspro.p1.generic-ppi-source@0.25.m2.score\endcsname{$+$84.7}
\expandafter\def\csname val@rev.pol.rules.sspro.p1.keep-all.keep\endcsname{1.000}
\expandafter\def\csname cnt@rev.pol.rules.sspro.p1.keep-all.keep\endcsname{1}
\expandafter\def\csname pct@rev.pol.rules.sspro.p1.keep-all.keep\endcsname{100.0}
\expandafter\def\csname ppt@rev.pol.rules.sspro.p1.keep-all.keep\endcsname{$+$100.0}
\expandafter\def\csname val@rev.pol.rules.sspro.p1.keep-all.req\endcsname{2.716}
\expandafter\def\csname cnt@rev.pol.rules.sspro.p1.keep-all.req\endcsname{3}
\expandafter\def\csname pct@rev.pol.rules.sspro.p1.keep-all.req\endcsname{271.6}
\expandafter\def\csname ppt@rev.pol.rules.sspro.p1.keep-all.req\endcsname{$+$271.6}
\expandafter\def\csname val@rev.pol.rules.sspro.p1.keep-all.score\endcsname{0.859}
\expandafter\def\csname cnt@rev.pol.rules.sspro.p1.keep-all.score\endcsname{1}
\expandafter\def\csname pct@rev.pol.rules.sspro.p1.keep-all.score\endcsname{85.9}
\expandafter\def\csname ppt@rev.pol.rules.sspro.p1.keep-all.score\endcsname{$+$85.9}
\expandafter\def\csname val@rev.pol.rules.sspro.p1.old-decision.frozen.keep\endcsname{1.000}
\expandafter\def\csname cnt@rev.pol.rules.sspro.p1.old-decision.frozen.keep\endcsname{1}
\expandafter\def\csname pct@rev.pol.rules.sspro.p1.old-decision.frozen.keep\endcsname{100.0}
\expandafter\def\csname ppt@rev.pol.rules.sspro.p1.old-decision.frozen.keep\endcsname{$+$100.0}
\expandafter\def\csname val@rev.pol.rules.sspro.p1.old-decision.frozen.req\endcsname{2.716}
\expandafter\def\csname cnt@rev.pol.rules.sspro.p1.old-decision.frozen.req\endcsname{3}
\expandafter\def\csname pct@rev.pol.rules.sspro.p1.old-decision.frozen.req\endcsname{271.6}
\expandafter\def\csname ppt@rev.pol.rules.sspro.p1.old-decision.frozen.req\endcsname{$+$271.6}
\expandafter\def\csname val@rev.pol.rules.sspro.p1.old-decision.frozen.score\endcsname{0.859}
\expandafter\def\csname cnt@rev.pol.rules.sspro.p1.old-decision.frozen.score\endcsname{1}
\expandafter\def\csname pct@rev.pol.rules.sspro.p1.old-decision.frozen.score\endcsname{85.9}
\expandafter\def\csname ppt@rev.pol.rules.sspro.p1.old-decision.frozen.score\endcsname{$+$85.9}
\expandafter\def\csname val@rev.pol.rules.sspro.p1.old-decision.lam0.5.keep\endcsname{1.000}
\expandafter\def\csname cnt@rev.pol.rules.sspro.p1.old-decision.lam0.5.keep\endcsname{1}
\expandafter\def\csname pct@rev.pol.rules.sspro.p1.old-decision.lam0.5.keep\endcsname{100.0}
\expandafter\def\csname ppt@rev.pol.rules.sspro.p1.old-decision.lam0.5.keep\endcsname{$+$100.0}
\expandafter\def\csname val@rev.pol.rules.sspro.p1.old-decision.lam0.5.req\endcsname{2.716}
\expandafter\def\csname cnt@rev.pol.rules.sspro.p1.old-decision.lam0.5.req\endcsname{3}
\expandafter\def\csname pct@rev.pol.rules.sspro.p1.old-decision.lam0.5.req\endcsname{271.6}
\expandafter\def\csname ppt@rev.pol.rules.sspro.p1.old-decision.lam0.5.req\endcsname{$+$271.6}
\expandafter\def\csname val@rev.pol.rules.sspro.p1.old-decision.lam0.5.score\endcsname{0.859}
\expandafter\def\csname cnt@rev.pol.rules.sspro.p1.old-decision.lam0.5.score\endcsname{1}
\expandafter\def\csname pct@rev.pol.rules.sspro.p1.old-decision.lam0.5.score\endcsname{85.9}
\expandafter\def\csname ppt@rev.pol.rules.sspro.p1.old-decision.lam0.5.score\endcsname{$+$85.9}
\expandafter\def\csname val@rev.pol.rules.sspro.p1.old-decision.lam1.keep\endcsname{1.000}
\expandafter\def\csname cnt@rev.pol.rules.sspro.p1.old-decision.lam1.keep\endcsname{1}
\expandafter\def\csname pct@rev.pol.rules.sspro.p1.old-decision.lam1.keep\endcsname{100.0}
\expandafter\def\csname ppt@rev.pol.rules.sspro.p1.old-decision.lam1.keep\endcsname{$+$100.0}
\expandafter\def\csname val@rev.pol.rules.sspro.p1.old-decision.lam1.req\endcsname{2.716}
\expandafter\def\csname cnt@rev.pol.rules.sspro.p1.old-decision.lam1.req\endcsname{3}
\expandafter\def\csname pct@rev.pol.rules.sspro.p1.old-decision.lam1.req\endcsname{271.6}
\expandafter\def\csname ppt@rev.pol.rules.sspro.p1.old-decision.lam1.req\endcsname{$+$271.6}
\expandafter\def\csname val@rev.pol.rules.sspro.p1.old-decision.lam1.score\endcsname{0.859}
\expandafter\def\csname cnt@rev.pol.rules.sspro.p1.old-decision.lam1.score\endcsname{1}
\expandafter\def\csname pct@rev.pol.rules.sspro.p1.old-decision.lam1.score\endcsname{85.9}
\expandafter\def\csname ppt@rev.pol.rules.sspro.p1.old-decision.lam1.score\endcsname{$+$85.9}
\expandafter\def\csname val@rev.pol.rules.sspro.p1.old-decision.m0.5.keep\endcsname{1.000}
\expandafter\def\csname cnt@rev.pol.rules.sspro.p1.old-decision.m0.5.keep\endcsname{1}
\expandafter\def\csname pct@rev.pol.rules.sspro.p1.old-decision.m0.5.keep\endcsname{100.0}
\expandafter\def\csname ppt@rev.pol.rules.sspro.p1.old-decision.m0.5.keep\endcsname{$+$100.0}
\expandafter\def\csname val@rev.pol.rules.sspro.p1.old-decision.m0.5.req\endcsname{2.716}
\expandafter\def\csname cnt@rev.pol.rules.sspro.p1.old-decision.m0.5.req\endcsname{3}
\expandafter\def\csname pct@rev.pol.rules.sspro.p1.old-decision.m0.5.req\endcsname{271.6}
\expandafter\def\csname ppt@rev.pol.rules.sspro.p1.old-decision.m0.5.req\endcsname{$+$271.6}
\expandafter\def\csname val@rev.pol.rules.sspro.p1.old-decision.m0.5.score\endcsname{0.859}
\expandafter\def\csname cnt@rev.pol.rules.sspro.p1.old-decision.m0.5.score\endcsname{1}
\expandafter\def\csname pct@rev.pol.rules.sspro.p1.old-decision.m0.5.score\endcsname{85.9}
\expandafter\def\csname ppt@rev.pol.rules.sspro.p1.old-decision.m0.5.score\endcsname{$+$85.9}
\expandafter\def\csname val@rev.pol.rules.sspro.p1.old-decision.m1.keep\endcsname{1.000}
\expandafter\def\csname cnt@rev.pol.rules.sspro.p1.old-decision.m1.keep\endcsname{1}
\expandafter\def\csname pct@rev.pol.rules.sspro.p1.old-decision.m1.keep\endcsname{100.0}
\expandafter\def\csname ppt@rev.pol.rules.sspro.p1.old-decision.m1.keep\endcsname{$+$100.0}
\expandafter\def\csname val@rev.pol.rules.sspro.p1.old-decision.m1.req\endcsname{2.716}
\expandafter\def\csname cnt@rev.pol.rules.sspro.p1.old-decision.m1.req\endcsname{3}
\expandafter\def\csname pct@rev.pol.rules.sspro.p1.old-decision.m1.req\endcsname{271.6}
\expandafter\def\csname ppt@rev.pol.rules.sspro.p1.old-decision.m1.req\endcsname{$+$271.6}
\expandafter\def\csname val@rev.pol.rules.sspro.p1.old-decision.m1.score\endcsname{0.859}
\expandafter\def\csname cnt@rev.pol.rules.sspro.p1.old-decision.m1.score\endcsname{1}
\expandafter\def\csname pct@rev.pol.rules.sspro.p1.old-decision.m1.score\endcsname{85.9}
\expandafter\def\csname ppt@rev.pol.rules.sspro.p1.old-decision.m1.score\endcsname{$+$85.9}
\expandafter\def\csname val@rev.pol.rules.sspro.p1.old-decision.m2.keep\endcsname{1.000}
\expandafter\def\csname cnt@rev.pol.rules.sspro.p1.old-decision.m2.keep\endcsname{1}
\expandafter\def\csname pct@rev.pol.rules.sspro.p1.old-decision.m2.keep\endcsname{100.0}
\expandafter\def\csname ppt@rev.pol.rules.sspro.p1.old-decision.m2.keep\endcsname{$+$100.0}
\expandafter\def\csname val@rev.pol.rules.sspro.p1.old-decision.m2.req\endcsname{2.716}
\expandafter\def\csname cnt@rev.pol.rules.sspro.p1.old-decision.m2.req\endcsname{3}
\expandafter\def\csname pct@rev.pol.rules.sspro.p1.old-decision.m2.req\endcsname{271.6}
\expandafter\def\csname ppt@rev.pol.rules.sspro.p1.old-decision.m2.req\endcsname{$+$271.6}
\expandafter\def\csname val@rev.pol.rules.sspro.p1.old-decision.m2.score\endcsname{0.859}
\expandafter\def\csname cnt@rev.pol.rules.sspro.p1.old-decision.m2.score\endcsname{1}
\expandafter\def\csname pct@rev.pol.rules.sspro.p1.old-decision.m2.score\endcsname{85.9}
\expandafter\def\csname ppt@rev.pol.rules.sspro.p1.old-decision.m2.score\endcsname{$+$85.9}
\expandafter\def\csname val@rev.pol.rules.sspro.p1.oracle-on-T.keep\endcsname{1.000}
\expandafter\def\csname cnt@rev.pol.rules.sspro.p1.oracle-on-T.keep\endcsname{1}
\expandafter\def\csname pct@rev.pol.rules.sspro.p1.oracle-on-T.keep\endcsname{100.0}
\expandafter\def\csname ppt@rev.pol.rules.sspro.p1.oracle-on-T.keep\endcsname{$+$100.0}
\expandafter\def\csname val@rev.pol.rules.sspro.p1.oracle-on-T.req\endcsname{2.716}
\expandafter\def\csname cnt@rev.pol.rules.sspro.p1.oracle-on-T.req\endcsname{3}
\expandafter\def\csname pct@rev.pol.rules.sspro.p1.oracle-on-T.req\endcsname{271.6}
\expandafter\def\csname ppt@rev.pol.rules.sspro.p1.oracle-on-T.req\endcsname{$+$271.6}
\expandafter\def\csname val@rev.pol.rules.sspro.p1.oracle-on-T.score\endcsname{0.859}
\expandafter\def\csname cnt@rev.pol.rules.sspro.p1.oracle-on-T.score\endcsname{1}
\expandafter\def\csname pct@rev.pol.rules.sspro.p1.oracle-on-T.score\endcsname{85.9}
\expandafter\def\csname ppt@rev.pol.rules.sspro.p1.oracle-on-T.score\endcsname{$+$85.9}
\expandafter\def\csname val@rev.pol.rules.sspro.p1.r-s.center-d1\endcsname{2.000}
\expandafter\def\csname cnt@rev.pol.rules.sspro.p1.r-s.center-d1\endcsname{2}
\expandafter\def\csname pct@rev.pol.rules.sspro.p1.r-s.center-d1\endcsname{200.0}
\expandafter\def\csname ppt@rev.pol.rules.sspro.p1.r-s.center-d1\endcsname{$+$200.0}
\expandafter\def\csname val@rev.pol.rules.sspro.p1.r-s.context-d1\endcsname{2.000}
\expandafter\def\csname cnt@rev.pol.rules.sspro.p1.r-s.context-d1\endcsname{2}
\expandafter\def\csname pct@rev.pol.rules.sspro.p1.r-s.context-d1\endcsname{200.0}
\expandafter\def\csname ppt@rev.pol.rules.sspro.p1.r-s.context-d1\endcsname{$+$200.0}
\expandafter\def\csname val@rev.pol.rules.sspro.p1.r-s.dual-d2\endcsname{3.021}
\expandafter\def\csname cnt@rev.pol.rules.sspro.p1.r-s.dual-d2\endcsname{3}
\expandafter\def\csname pct@rev.pol.rules.sspro.p1.r-s.dual-d2\endcsname{302.1}
\expandafter\def\csname ppt@rev.pol.rules.sspro.p1.r-s.dual-d2\endcsname{$+$302.1}
\expandafter\def\csname val@rev.pol.rules.sspro.p1.r-s.grid2-d3\endcsname{3.938}
\expandafter\def\csname cnt@rev.pol.rules.sspro.p1.r-s.grid2-d3\endcsname{4}
\expandafter\def\csname pct@rev.pol.rules.sspro.p1.r-s.grid2-d3\endcsname{393.8}
\expandafter\def\csname ppt@rev.pol.rules.sspro.p1.r-s.grid2-d3\endcsname{$+$393.8}
\expandafter\def\csname val@rev.pol.rules.sspro.p1.rcr-stratified@0.25.frozen.keep\endcsname{0.973}
\expandafter\def\csname cnt@rev.pol.rules.sspro.p1.rcr-stratified@0.25.frozen.keep\endcsname{1}
\expandafter\def\csname pct@rev.pol.rules.sspro.p1.rcr-stratified@0.25.frozen.keep\endcsname{97.3}
\expandafter\def\csname ppt@rev.pol.rules.sspro.p1.rcr-stratified@0.25.frozen.keep\endcsname{$+$97.3}
\expandafter\def\csname val@rev.pol.rules.sspro.p1.rcr-stratified@0.25.frozen.req\endcsname{2.658}
\expandafter\def\csname cnt@rev.pol.rules.sspro.p1.rcr-stratified@0.25.frozen.req\endcsname{3}
\expandafter\def\csname pct@rev.pol.rules.sspro.p1.rcr-stratified@0.25.frozen.req\endcsname{265.8}
\expandafter\def\csname ppt@rev.pol.rules.sspro.p1.rcr-stratified@0.25.frozen.req\endcsname{$+$265.8}
\expandafter\def\csname val@rev.pol.rules.sspro.p1.rcr-stratified@0.25.frozen.score\endcsname{0.856}
\expandafter\def\csname cnt@rev.pol.rules.sspro.p1.rcr-stratified@0.25.frozen.score\endcsname{1}
\expandafter\def\csname pct@rev.pol.rules.sspro.p1.rcr-stratified@0.25.frozen.score\endcsname{85.6}
\expandafter\def\csname ppt@rev.pol.rules.sspro.p1.rcr-stratified@0.25.frozen.score\endcsname{$+$85.6}
\expandafter\def\csname val@rev.pol.rules.sspro.p1.rcr-stratified@0.25.lam0.5.keep\endcsname{0.944}
\expandafter\def\csname cnt@rev.pol.rules.sspro.p1.rcr-stratified@0.25.lam0.5.keep\endcsname{1}
\expandafter\def\csname pct@rev.pol.rules.sspro.p1.rcr-stratified@0.25.lam0.5.keep\endcsname{94.3}
\expandafter\def\csname ppt@rev.pol.rules.sspro.p1.rcr-stratified@0.25.lam0.5.keep\endcsname{$+$94.3}
\expandafter\def\csname val@rev.pol.rules.sspro.p1.rcr-stratified@0.25.lam0.5.req\endcsname{2.597}
\expandafter\def\csname cnt@rev.pol.rules.sspro.p1.rcr-stratified@0.25.lam0.5.req\endcsname{3}
\expandafter\def\csname pct@rev.pol.rules.sspro.p1.rcr-stratified@0.25.lam0.5.req\endcsname{259.7}
\expandafter\def\csname ppt@rev.pol.rules.sspro.p1.rcr-stratified@0.25.lam0.5.req\endcsname{$+$259.7}
\expandafter\def\csname val@rev.pol.rules.sspro.p1.rcr-stratified@0.25.lam0.5.score\endcsname{0.851}
\expandafter\def\csname cnt@rev.pol.rules.sspro.p1.rcr-stratified@0.25.lam0.5.score\endcsname{1}
\expandafter\def\csname pct@rev.pol.rules.sspro.p1.rcr-stratified@0.25.lam0.5.score\endcsname{85.1}
\expandafter\def\csname ppt@rev.pol.rules.sspro.p1.rcr-stratified@0.25.lam0.5.score\endcsname{$+$85.1}
\expandafter\def\csname val@rev.pol.rules.sspro.p1.rcr-stratified@0.25.lam1.keep\endcsname{0.915}
\expandafter\def\csname cnt@rev.pol.rules.sspro.p1.rcr-stratified@0.25.lam1.keep\endcsname{1}
\expandafter\def\csname pct@rev.pol.rules.sspro.p1.rcr-stratified@0.25.lam1.keep\endcsname{91.5}
\expandafter\def\csname ppt@rev.pol.rules.sspro.p1.rcr-stratified@0.25.lam1.keep\endcsname{$+$91.5}
\expandafter\def\csname val@rev.pol.rules.sspro.p1.rcr-stratified@0.25.lam1.req\endcsname{2.517}
\expandafter\def\csname cnt@rev.pol.rules.sspro.p1.rcr-stratified@0.25.lam1.req\endcsname{3}
\expandafter\def\csname pct@rev.pol.rules.sspro.p1.rcr-stratified@0.25.lam1.req\endcsname{251.7}
\expandafter\def\csname ppt@rev.pol.rules.sspro.p1.rcr-stratified@0.25.lam1.req\endcsname{$+$251.7}
\expandafter\def\csname val@rev.pol.rules.sspro.p1.rcr-stratified@0.25.lam1.score\endcsname{0.848}
\expandafter\def\csname cnt@rev.pol.rules.sspro.p1.rcr-stratified@0.25.lam1.score\endcsname{1}
\expandafter\def\csname pct@rev.pol.rules.sspro.p1.rcr-stratified@0.25.lam1.score\endcsname{84.8}
\expandafter\def\csname ppt@rev.pol.rules.sspro.p1.rcr-stratified@0.25.lam1.score\endcsname{$+$84.8}
\expandafter\def\csname val@rev.pol.rules.sspro.p1.rcr-stratified@0.25.m0.5.keep\endcsname{0.973}
\expandafter\def\csname cnt@rev.pol.rules.sspro.p1.rcr-stratified@0.25.m0.5.keep\endcsname{1}
\expandafter\def\csname pct@rev.pol.rules.sspro.p1.rcr-stratified@0.25.m0.5.keep\endcsname{97.3}
\expandafter\def\csname ppt@rev.pol.rules.sspro.p1.rcr-stratified@0.25.m0.5.keep\endcsname{$+$97.3}
\expandafter\def\csname val@rev.pol.rules.sspro.p1.rcr-stratified@0.25.m0.5.req\endcsname{2.658}
\expandafter\def\csname cnt@rev.pol.rules.sspro.p1.rcr-stratified@0.25.m0.5.req\endcsname{3}
\expandafter\def\csname pct@rev.pol.rules.sspro.p1.rcr-stratified@0.25.m0.5.req\endcsname{265.8}
\expandafter\def\csname ppt@rev.pol.rules.sspro.p1.rcr-stratified@0.25.m0.5.req\endcsname{$+$265.8}
\expandafter\def\csname val@rev.pol.rules.sspro.p1.rcr-stratified@0.25.m0.5.score\endcsname{0.856}
\expandafter\def\csname cnt@rev.pol.rules.sspro.p1.rcr-stratified@0.25.m0.5.score\endcsname{1}
\expandafter\def\csname pct@rev.pol.rules.sspro.p1.rcr-stratified@0.25.m0.5.score\endcsname{85.6}
\expandafter\def\csname ppt@rev.pol.rules.sspro.p1.rcr-stratified@0.25.m0.5.score\endcsname{$+$85.6}
\expandafter\def\csname val@rev.pol.rules.sspro.p1.rcr-stratified@0.25.m1.keep\endcsname{0.944}
\expandafter\def\csname cnt@rev.pol.rules.sspro.p1.rcr-stratified@0.25.m1.keep\endcsname{1}
\expandafter\def\csname pct@rev.pol.rules.sspro.p1.rcr-stratified@0.25.m1.keep\endcsname{94.3}
\expandafter\def\csname ppt@rev.pol.rules.sspro.p1.rcr-stratified@0.25.m1.keep\endcsname{$+$94.3}
\expandafter\def\csname val@rev.pol.rules.sspro.p1.rcr-stratified@0.25.m1.req\endcsname{2.597}
\expandafter\def\csname cnt@rev.pol.rules.sspro.p1.rcr-stratified@0.25.m1.req\endcsname{3}
\expandafter\def\csname pct@rev.pol.rules.sspro.p1.rcr-stratified@0.25.m1.req\endcsname{259.7}
\expandafter\def\csname ppt@rev.pol.rules.sspro.p1.rcr-stratified@0.25.m1.req\endcsname{$+$259.7}
\expandafter\def\csname val@rev.pol.rules.sspro.p1.rcr-stratified@0.25.m1.score\endcsname{0.851}
\expandafter\def\csname cnt@rev.pol.rules.sspro.p1.rcr-stratified@0.25.m1.score\endcsname{1}
\expandafter\def\csname pct@rev.pol.rules.sspro.p1.rcr-stratified@0.25.m1.score\endcsname{85.1}
\expandafter\def\csname ppt@rev.pol.rules.sspro.p1.rcr-stratified@0.25.m1.score\endcsname{$+$85.1}
\expandafter\def\csname val@rev.pol.rules.sspro.p1.rcr-stratified@0.25.m2.keep\endcsname{0.943}
\expandafter\def\csname cnt@rev.pol.rules.sspro.p1.rcr-stratified@0.25.m2.keep\endcsname{1}
\expandafter\def\csname pct@rev.pol.rules.sspro.p1.rcr-stratified@0.25.m2.keep\endcsname{94.3}
\expandafter\def\csname ppt@rev.pol.rules.sspro.p1.rcr-stratified@0.25.m2.keep\endcsname{$+$94.3}
\expandafter\def\csname val@rev.pol.rules.sspro.p1.rcr-stratified@0.25.m2.req\endcsname{2.595}
\expandafter\def\csname cnt@rev.pol.rules.sspro.p1.rcr-stratified@0.25.m2.req\endcsname{3}
\expandafter\def\csname pct@rev.pol.rules.sspro.p1.rcr-stratified@0.25.m2.req\endcsname{259.5}
\expandafter\def\csname ppt@rev.pol.rules.sspro.p1.rcr-stratified@0.25.m2.req\endcsname{$+$259.5}
\expandafter\def\csname val@rev.pol.rules.sspro.p1.rcr-stratified@0.25.m2.score\endcsname{0.851}
\expandafter\def\csname cnt@rev.pol.rules.sspro.p1.rcr-stratified@0.25.m2.score\endcsname{1}
\expandafter\def\csname pct@rev.pol.rules.sspro.p1.rcr-stratified@0.25.m2.score\endcsname{85.1}
\expandafter\def\csname ppt@rev.pol.rules.sspro.p1.rcr-stratified@0.25.m2.score\endcsname{$+$85.1}
\expandafter\def\csname val@rev.pol.rules.sspro.p1.replay-only.frozen.keep\endcsname{1.000}
\expandafter\def\csname cnt@rev.pol.rules.sspro.p1.replay-only.frozen.keep\endcsname{1}
\expandafter\def\csname pct@rev.pol.rules.sspro.p1.replay-only.frozen.keep\endcsname{100.0}
\expandafter\def\csname ppt@rev.pol.rules.sspro.p1.replay-only.frozen.keep\endcsname{$+$100.0}
\expandafter\def\csname val@rev.pol.rules.sspro.p1.replay-only.frozen.req\endcsname{2.716}
\expandafter\def\csname cnt@rev.pol.rules.sspro.p1.replay-only.frozen.req\endcsname{3}
\expandafter\def\csname pct@rev.pol.rules.sspro.p1.replay-only.frozen.req\endcsname{271.6}
\expandafter\def\csname ppt@rev.pol.rules.sspro.p1.replay-only.frozen.req\endcsname{$+$271.6}
\expandafter\def\csname val@rev.pol.rules.sspro.p1.replay-only.frozen.score\endcsname{0.859}
\expandafter\def\csname cnt@rev.pol.rules.sspro.p1.replay-only.frozen.score\endcsname{1}
\expandafter\def\csname pct@rev.pol.rules.sspro.p1.replay-only.frozen.score\endcsname{85.9}
\expandafter\def\csname ppt@rev.pol.rules.sspro.p1.replay-only.frozen.score\endcsname{$+$85.9}
\expandafter\def\csname val@rev.pol.rules.sspro.p1.replay-only.lam0.5.keep\endcsname{1.000}
\expandafter\def\csname cnt@rev.pol.rules.sspro.p1.replay-only.lam0.5.keep\endcsname{1}
\expandafter\def\csname pct@rev.pol.rules.sspro.p1.replay-only.lam0.5.keep\endcsname{100.0}
\expandafter\def\csname ppt@rev.pol.rules.sspro.p1.replay-only.lam0.5.keep\endcsname{$+$100.0}
\expandafter\def\csname val@rev.pol.rules.sspro.p1.replay-only.lam0.5.req\endcsname{2.716}
\expandafter\def\csname cnt@rev.pol.rules.sspro.p1.replay-only.lam0.5.req\endcsname{3}
\expandafter\def\csname pct@rev.pol.rules.sspro.p1.replay-only.lam0.5.req\endcsname{271.6}
\expandafter\def\csname ppt@rev.pol.rules.sspro.p1.replay-only.lam0.5.req\endcsname{$+$271.6}
\expandafter\def\csname val@rev.pol.rules.sspro.p1.replay-only.lam0.5.score\endcsname{0.859}
\expandafter\def\csname cnt@rev.pol.rules.sspro.p1.replay-only.lam0.5.score\endcsname{1}
\expandafter\def\csname pct@rev.pol.rules.sspro.p1.replay-only.lam0.5.score\endcsname{85.9}
\expandafter\def\csname ppt@rev.pol.rules.sspro.p1.replay-only.lam0.5.score\endcsname{$+$85.9}
\expandafter\def\csname val@rev.pol.rules.sspro.p1.replay-only.lam1.keep\endcsname{1.000}
\expandafter\def\csname cnt@rev.pol.rules.sspro.p1.replay-only.lam1.keep\endcsname{1}
\expandafter\def\csname pct@rev.pol.rules.sspro.p1.replay-only.lam1.keep\endcsname{100.0}
\expandafter\def\csname ppt@rev.pol.rules.sspro.p1.replay-only.lam1.keep\endcsname{$+$100.0}
\expandafter\def\csname val@rev.pol.rules.sspro.p1.replay-only.lam1.req\endcsname{2.716}
\expandafter\def\csname cnt@rev.pol.rules.sspro.p1.replay-only.lam1.req\endcsname{3}
\expandafter\def\csname pct@rev.pol.rules.sspro.p1.replay-only.lam1.req\endcsname{271.6}
\expandafter\def\csname ppt@rev.pol.rules.sspro.p1.replay-only.lam1.req\endcsname{$+$271.6}
\expandafter\def\csname val@rev.pol.rules.sspro.p1.replay-only.lam1.score\endcsname{0.859}
\expandafter\def\csname cnt@rev.pol.rules.sspro.p1.replay-only.lam1.score\endcsname{1}
\expandafter\def\csname pct@rev.pol.rules.sspro.p1.replay-only.lam1.score\endcsname{85.9}
\expandafter\def\csname ppt@rev.pol.rules.sspro.p1.replay-only.lam1.score\endcsname{$+$85.9}
\expandafter\def\csname val@rev.pol.rules.sspro.p1.replay-only.m0.5.keep\endcsname{1.000}
\expandafter\def\csname cnt@rev.pol.rules.sspro.p1.replay-only.m0.5.keep\endcsname{1}
\expandafter\def\csname pct@rev.pol.rules.sspro.p1.replay-only.m0.5.keep\endcsname{100.0}
\expandafter\def\csname ppt@rev.pol.rules.sspro.p1.replay-only.m0.5.keep\endcsname{$+$100.0}
\expandafter\def\csname val@rev.pol.rules.sspro.p1.replay-only.m0.5.req\endcsname{2.716}
\expandafter\def\csname cnt@rev.pol.rules.sspro.p1.replay-only.m0.5.req\endcsname{3}
\expandafter\def\csname pct@rev.pol.rules.sspro.p1.replay-only.m0.5.req\endcsname{271.6}
\expandafter\def\csname ppt@rev.pol.rules.sspro.p1.replay-only.m0.5.req\endcsname{$+$271.6}
\expandafter\def\csname val@rev.pol.rules.sspro.p1.replay-only.m0.5.score\endcsname{0.859}
\expandafter\def\csname cnt@rev.pol.rules.sspro.p1.replay-only.m0.5.score\endcsname{1}
\expandafter\def\csname pct@rev.pol.rules.sspro.p1.replay-only.m0.5.score\endcsname{85.9}
\expandafter\def\csname ppt@rev.pol.rules.sspro.p1.replay-only.m0.5.score\endcsname{$+$85.9}
\expandafter\def\csname val@rev.pol.rules.sspro.p1.replay-only.m1.keep\endcsname{1.000}
\expandafter\def\csname cnt@rev.pol.rules.sspro.p1.replay-only.m1.keep\endcsname{1}
\expandafter\def\csname pct@rev.pol.rules.sspro.p1.replay-only.m1.keep\endcsname{100.0}
\expandafter\def\csname ppt@rev.pol.rules.sspro.p1.replay-only.m1.keep\endcsname{$+$100.0}
\expandafter\def\csname val@rev.pol.rules.sspro.p1.replay-only.m1.req\endcsname{2.716}
\expandafter\def\csname cnt@rev.pol.rules.sspro.p1.replay-only.m1.req\endcsname{3}
\expandafter\def\csname pct@rev.pol.rules.sspro.p1.replay-only.m1.req\endcsname{271.6}
\expandafter\def\csname ppt@rev.pol.rules.sspro.p1.replay-only.m1.req\endcsname{$+$271.6}
\expandafter\def\csname val@rev.pol.rules.sspro.p1.replay-only.m1.score\endcsname{0.859}
\expandafter\def\csname cnt@rev.pol.rules.sspro.p1.replay-only.m1.score\endcsname{1}
\expandafter\def\csname pct@rev.pol.rules.sspro.p1.replay-only.m1.score\endcsname{85.9}
\expandafter\def\csname ppt@rev.pol.rules.sspro.p1.replay-only.m1.score\endcsname{$+$85.9}
\expandafter\def\csname val@rev.pol.rules.sspro.p1.replay-only.m2.keep\endcsname{1.000}
\expandafter\def\csname cnt@rev.pol.rules.sspro.p1.replay-only.m2.keep\endcsname{1}
\expandafter\def\csname pct@rev.pol.rules.sspro.p1.replay-only.m2.keep\endcsname{100.0}
\expandafter\def\csname ppt@rev.pol.rules.sspro.p1.replay-only.m2.keep\endcsname{$+$100.0}
\expandafter\def\csname val@rev.pol.rules.sspro.p1.replay-only.m2.req\endcsname{2.716}
\expandafter\def\csname cnt@rev.pol.rules.sspro.p1.replay-only.m2.req\endcsname{3}
\expandafter\def\csname pct@rev.pol.rules.sspro.p1.replay-only.m2.req\endcsname{271.6}
\expandafter\def\csname ppt@rev.pol.rules.sspro.p1.replay-only.m2.req\endcsname{$+$271.6}
\expandafter\def\csname val@rev.pol.rules.sspro.p1.replay-only.m2.score\endcsname{0.859}
\expandafter\def\csname cnt@rev.pol.rules.sspro.p1.replay-only.m2.score\endcsname{1}
\expandafter\def\csname pct@rev.pol.rules.sspro.p1.replay-only.m2.score\endcsname{85.9}
\expandafter\def\csname ppt@rev.pol.rules.sspro.p1.replay-only.m2.score\endcsname{$+$85.9}
\expandafter\def\csname val@rev.pol.rules.sspro.p1.replay-ppi-uniform@0.25.frozen.keep\endcsname{0.949}
\expandafter\def\csname cnt@rev.pol.rules.sspro.p1.replay-ppi-uniform@0.25.frozen.keep\endcsname{1}
\expandafter\def\csname pct@rev.pol.rules.sspro.p1.replay-ppi-uniform@0.25.frozen.keep\endcsname{94.9}
\expandafter\def\csname ppt@rev.pol.rules.sspro.p1.replay-ppi-uniform@0.25.frozen.keep\endcsname{$+$94.9}
\expandafter\def\csname val@rev.pol.rules.sspro.p1.replay-ppi-uniform@0.25.frozen.req\endcsname{2.619}
\expandafter\def\csname cnt@rev.pol.rules.sspro.p1.replay-ppi-uniform@0.25.frozen.req\endcsname{3}
\expandafter\def\csname pct@rev.pol.rules.sspro.p1.replay-ppi-uniform@0.25.frozen.req\endcsname{261.9}
\expandafter\def\csname ppt@rev.pol.rules.sspro.p1.replay-ppi-uniform@0.25.frozen.req\endcsname{$+$261.9}
\expandafter\def\csname val@rev.pol.rules.sspro.p1.replay-ppi-uniform@0.25.frozen.score\endcsname{0.852}
\expandafter\def\csname cnt@rev.pol.rules.sspro.p1.replay-ppi-uniform@0.25.frozen.score\endcsname{1}
\expandafter\def\csname pct@rev.pol.rules.sspro.p1.replay-ppi-uniform@0.25.frozen.score\endcsname{85.2}
\expandafter\def\csname ppt@rev.pol.rules.sspro.p1.replay-ppi-uniform@0.25.frozen.score\endcsname{$+$85.2}
\expandafter\def\csname val@rev.pol.rules.sspro.p1.replay-ppi-uniform@0.25.lam0.5.keep\endcsname{0.949}
\expandafter\def\csname cnt@rev.pol.rules.sspro.p1.replay-ppi-uniform@0.25.lam0.5.keep\endcsname{1}
\expandafter\def\csname pct@rev.pol.rules.sspro.p1.replay-ppi-uniform@0.25.lam0.5.keep\endcsname{94.9}
\expandafter\def\csname ppt@rev.pol.rules.sspro.p1.replay-ppi-uniform@0.25.lam0.5.keep\endcsname{$+$94.9}
\expandafter\def\csname val@rev.pol.rules.sspro.p1.replay-ppi-uniform@0.25.lam0.5.req\endcsname{2.619}
\expandafter\def\csname cnt@rev.pol.rules.sspro.p1.replay-ppi-uniform@0.25.lam0.5.req\endcsname{3}
\expandafter\def\csname pct@rev.pol.rules.sspro.p1.replay-ppi-uniform@0.25.lam0.5.req\endcsname{261.9}
\expandafter\def\csname ppt@rev.pol.rules.sspro.p1.replay-ppi-uniform@0.25.lam0.5.req\endcsname{$+$261.9}
\expandafter\def\csname val@rev.pol.rules.sspro.p1.replay-ppi-uniform@0.25.lam0.5.score\endcsname{0.852}
\expandafter\def\csname cnt@rev.pol.rules.sspro.p1.replay-ppi-uniform@0.25.lam0.5.score\endcsname{1}
\expandafter\def\csname pct@rev.pol.rules.sspro.p1.replay-ppi-uniform@0.25.lam0.5.score\endcsname{85.2}
\expandafter\def\csname ppt@rev.pol.rules.sspro.p1.replay-ppi-uniform@0.25.lam0.5.score\endcsname{$+$85.2}
\expandafter\def\csname val@rev.pol.rules.sspro.p1.replay-ppi-uniform@0.25.lam1.keep\endcsname{0.949}
\expandafter\def\csname cnt@rev.pol.rules.sspro.p1.replay-ppi-uniform@0.25.lam1.keep\endcsname{1}
\expandafter\def\csname pct@rev.pol.rules.sspro.p1.replay-ppi-uniform@0.25.lam1.keep\endcsname{94.9}
\expandafter\def\csname ppt@rev.pol.rules.sspro.p1.replay-ppi-uniform@0.25.lam1.keep\endcsname{$+$94.9}
\expandafter\def\csname val@rev.pol.rules.sspro.p1.replay-ppi-uniform@0.25.lam1.req\endcsname{2.619}
\expandafter\def\csname cnt@rev.pol.rules.sspro.p1.replay-ppi-uniform@0.25.lam1.req\endcsname{3}
\expandafter\def\csname pct@rev.pol.rules.sspro.p1.replay-ppi-uniform@0.25.lam1.req\endcsname{261.9}
\expandafter\def\csname ppt@rev.pol.rules.sspro.p1.replay-ppi-uniform@0.25.lam1.req\endcsname{$+$261.9}
\expandafter\def\csname val@rev.pol.rules.sspro.p1.replay-ppi-uniform@0.25.lam1.score\endcsname{0.852}
\expandafter\def\csname cnt@rev.pol.rules.sspro.p1.replay-ppi-uniform@0.25.lam1.score\endcsname{1}
\expandafter\def\csname pct@rev.pol.rules.sspro.p1.replay-ppi-uniform@0.25.lam1.score\endcsname{85.2}
\expandafter\def\csname ppt@rev.pol.rules.sspro.p1.replay-ppi-uniform@0.25.lam1.score\endcsname{$+$85.2}
\expandafter\def\csname val@rev.pol.rules.sspro.p1.replay-ppi-uniform@0.25.m0.5.keep\endcsname{0.949}
\expandafter\def\csname cnt@rev.pol.rules.sspro.p1.replay-ppi-uniform@0.25.m0.5.keep\endcsname{1}
\expandafter\def\csname pct@rev.pol.rules.sspro.p1.replay-ppi-uniform@0.25.m0.5.keep\endcsname{94.9}
\expandafter\def\csname ppt@rev.pol.rules.sspro.p1.replay-ppi-uniform@0.25.m0.5.keep\endcsname{$+$94.9}
\expandafter\def\csname val@rev.pol.rules.sspro.p1.replay-ppi-uniform@0.25.m0.5.req\endcsname{2.619}
\expandafter\def\csname cnt@rev.pol.rules.sspro.p1.replay-ppi-uniform@0.25.m0.5.req\endcsname{3}
\expandafter\def\csname pct@rev.pol.rules.sspro.p1.replay-ppi-uniform@0.25.m0.5.req\endcsname{261.9}
\expandafter\def\csname ppt@rev.pol.rules.sspro.p1.replay-ppi-uniform@0.25.m0.5.req\endcsname{$+$261.9}
\expandafter\def\csname val@rev.pol.rules.sspro.p1.replay-ppi-uniform@0.25.m0.5.score\endcsname{0.852}
\expandafter\def\csname cnt@rev.pol.rules.sspro.p1.replay-ppi-uniform@0.25.m0.5.score\endcsname{1}
\expandafter\def\csname pct@rev.pol.rules.sspro.p1.replay-ppi-uniform@0.25.m0.5.score\endcsname{85.2}
\expandafter\def\csname ppt@rev.pol.rules.sspro.p1.replay-ppi-uniform@0.25.m0.5.score\endcsname{$+$85.2}
\expandafter\def\csname val@rev.pol.rules.sspro.p1.replay-ppi-uniform@0.25.m1.keep\endcsname{0.949}
\expandafter\def\csname cnt@rev.pol.rules.sspro.p1.replay-ppi-uniform@0.25.m1.keep\endcsname{1}
\expandafter\def\csname pct@rev.pol.rules.sspro.p1.replay-ppi-uniform@0.25.m1.keep\endcsname{94.9}
\expandafter\def\csname ppt@rev.pol.rules.sspro.p1.replay-ppi-uniform@0.25.m1.keep\endcsname{$+$94.9}
\expandafter\def\csname val@rev.pol.rules.sspro.p1.replay-ppi-uniform@0.25.m1.req\endcsname{2.619}
\expandafter\def\csname cnt@rev.pol.rules.sspro.p1.replay-ppi-uniform@0.25.m1.req\endcsname{3}
\expandafter\def\csname pct@rev.pol.rules.sspro.p1.replay-ppi-uniform@0.25.m1.req\endcsname{261.9}
\expandafter\def\csname ppt@rev.pol.rules.sspro.p1.replay-ppi-uniform@0.25.m1.req\endcsname{$+$261.9}
\expandafter\def\csname val@rev.pol.rules.sspro.p1.replay-ppi-uniform@0.25.m1.score\endcsname{0.852}
\expandafter\def\csname cnt@rev.pol.rules.sspro.p1.replay-ppi-uniform@0.25.m1.score\endcsname{1}
\expandafter\def\csname pct@rev.pol.rules.sspro.p1.replay-ppi-uniform@0.25.m1.score\endcsname{85.2}
\expandafter\def\csname ppt@rev.pol.rules.sspro.p1.replay-ppi-uniform@0.25.m1.score\endcsname{$+$85.2}
\expandafter\def\csname val@rev.pol.rules.sspro.p1.replay-ppi-uniform@0.25.m2.keep\endcsname{0.949}
\expandafter\def\csname cnt@rev.pol.rules.sspro.p1.replay-ppi-uniform@0.25.m2.keep\endcsname{1}
\expandafter\def\csname pct@rev.pol.rules.sspro.p1.replay-ppi-uniform@0.25.m2.keep\endcsname{94.9}
\expandafter\def\csname ppt@rev.pol.rules.sspro.p1.replay-ppi-uniform@0.25.m2.keep\endcsname{$+$94.9}
\expandafter\def\csname val@rev.pol.rules.sspro.p1.replay-ppi-uniform@0.25.m2.req\endcsname{2.619}
\expandafter\def\csname cnt@rev.pol.rules.sspro.p1.replay-ppi-uniform@0.25.m2.req\endcsname{3}
\expandafter\def\csname pct@rev.pol.rules.sspro.p1.replay-ppi-uniform@0.25.m2.req\endcsname{261.9}
\expandafter\def\csname ppt@rev.pol.rules.sspro.p1.replay-ppi-uniform@0.25.m2.req\endcsname{$+$261.9}
\expandafter\def\csname val@rev.pol.rules.sspro.p1.replay-ppi-uniform@0.25.m2.score\endcsname{0.852}
\expandafter\def\csname cnt@rev.pol.rules.sspro.p1.replay-ppi-uniform@0.25.m2.score\endcsname{1}
\expandafter\def\csname pct@rev.pol.rules.sspro.p1.replay-ppi-uniform@0.25.m2.score\endcsname{85.2}
\expandafter\def\csname ppt@rev.pol.rules.sspro.p1.replay-ppi-uniform@0.25.m2.score\endcsname{$+$85.2}
\expandafter\def\csname val@rev.pol.rules.sspro.p1.suspend-all.keep\endcsname{0.000}
\expandafter\def\csname cnt@rev.pol.rules.sspro.p1.suspend-all.keep\endcsname{0}
\expandafter\def\csname pct@rev.pol.rules.sspro.p1.suspend-all.keep\endcsname{0.0}
\expandafter\def\csname ppt@rev.pol.rules.sspro.p1.suspend-all.keep\endcsname{0.0}
\expandafter\def\csname val@rev.pol.rules.sspro.p1.suspend-all.req\endcsname{1.000}
\expandafter\def\csname cnt@rev.pol.rules.sspro.p1.suspend-all.req\endcsname{1}
\expandafter\def\csname pct@rev.pol.rules.sspro.p1.suspend-all.req\endcsname{100.0}
\expandafter\def\csname ppt@rev.pol.rules.sspro.p1.suspend-all.req\endcsname{$+$100.0}
\expandafter\def\csname val@rev.pol.rules.sspro.p1.suspend-all.score\endcsname{0.740}
\expandafter\def\csname cnt@rev.pol.rules.sspro.p1.suspend-all.score\endcsname{1}
\expandafter\def\csname pct@rev.pol.rules.sspro.p1.suspend-all.score\endcsname{74.0}
\expandafter\def\csname ppt@rev.pol.rules.sspro.p1.suspend-all.score\endcsname{$+$74.0}
\expandafter\def\csname val@rev.pol.rules.sspro.p1.uniform-fresh@0.25.frozen.keep\endcsname{0.949}
\expandafter\def\csname cnt@rev.pol.rules.sspro.p1.uniform-fresh@0.25.frozen.keep\endcsname{1}
\expandafter\def\csname pct@rev.pol.rules.sspro.p1.uniform-fresh@0.25.frozen.keep\endcsname{94.9}
\expandafter\def\csname ppt@rev.pol.rules.sspro.p1.uniform-fresh@0.25.frozen.keep\endcsname{$+$94.9}
\expandafter\def\csname val@rev.pol.rules.sspro.p1.uniform-fresh@0.25.frozen.req\endcsname{2.629}
\expandafter\def\csname cnt@rev.pol.rules.sspro.p1.uniform-fresh@0.25.frozen.req\endcsname{3}
\expandafter\def\csname pct@rev.pol.rules.sspro.p1.uniform-fresh@0.25.frozen.req\endcsname{262.9}
\expandafter\def\csname ppt@rev.pol.rules.sspro.p1.uniform-fresh@0.25.frozen.req\endcsname{$+$262.9}
\expandafter\def\csname val@rev.pol.rules.sspro.p1.uniform-fresh@0.25.frozen.score\endcsname{0.853}
\expandafter\def\csname cnt@rev.pol.rules.sspro.p1.uniform-fresh@0.25.frozen.score\endcsname{1}
\expandafter\def\csname pct@rev.pol.rules.sspro.p1.uniform-fresh@0.25.frozen.score\endcsname{85.3}
\expandafter\def\csname ppt@rev.pol.rules.sspro.p1.uniform-fresh@0.25.frozen.score\endcsname{$+$85.3}
\expandafter\def\csname val@rev.pol.rules.sspro.p1.uniform-fresh@0.25.lam0.5.keep\endcsname{0.797}
\expandafter\def\csname cnt@rev.pol.rules.sspro.p1.uniform-fresh@0.25.lam0.5.keep\endcsname{1}
\expandafter\def\csname pct@rev.pol.rules.sspro.p1.uniform-fresh@0.25.lam0.5.keep\endcsname{79.7}
\expandafter\def\csname ppt@rev.pol.rules.sspro.p1.uniform-fresh@0.25.lam0.5.keep\endcsname{$+$79.7}
\expandafter\def\csname val@rev.pol.rules.sspro.p1.uniform-fresh@0.25.lam0.5.req\endcsname{2.372}
\expandafter\def\csname cnt@rev.pol.rules.sspro.p1.uniform-fresh@0.25.lam0.5.req\endcsname{2}
\expandafter\def\csname pct@rev.pol.rules.sspro.p1.uniform-fresh@0.25.lam0.5.req\endcsname{237.2}
\expandafter\def\csname ppt@rev.pol.rules.sspro.p1.uniform-fresh@0.25.lam0.5.req\endcsname{$+$237.2}
\expandafter\def\csname val@rev.pol.rules.sspro.p1.uniform-fresh@0.25.lam0.5.score\endcsname{0.835}
\expandafter\def\csname cnt@rev.pol.rules.sspro.p1.uniform-fresh@0.25.lam0.5.score\endcsname{1}
\expandafter\def\csname pct@rev.pol.rules.sspro.p1.uniform-fresh@0.25.lam0.5.score\endcsname{83.5}
\expandafter\def\csname ppt@rev.pol.rules.sspro.p1.uniform-fresh@0.25.lam0.5.score\endcsname{$+$83.5}
\expandafter\def\csname val@rev.pol.rules.sspro.p1.uniform-fresh@0.25.lam1.keep\endcsname{0.797}
\expandafter\def\csname cnt@rev.pol.rules.sspro.p1.uniform-fresh@0.25.lam1.keep\endcsname{1}
\expandafter\def\csname pct@rev.pol.rules.sspro.p1.uniform-fresh@0.25.lam1.keep\endcsname{79.7}
\expandafter\def\csname ppt@rev.pol.rules.sspro.p1.uniform-fresh@0.25.lam1.keep\endcsname{$+$79.7}
\expandafter\def\csname val@rev.pol.rules.sspro.p1.uniform-fresh@0.25.lam1.req\endcsname{2.372}
\expandafter\def\csname cnt@rev.pol.rules.sspro.p1.uniform-fresh@0.25.lam1.req\endcsname{2}
\expandafter\def\csname pct@rev.pol.rules.sspro.p1.uniform-fresh@0.25.lam1.req\endcsname{237.2}
\expandafter\def\csname ppt@rev.pol.rules.sspro.p1.uniform-fresh@0.25.lam1.req\endcsname{$+$237.2}
\expandafter\def\csname val@rev.pol.rules.sspro.p1.uniform-fresh@0.25.lam1.score\endcsname{0.835}
\expandafter\def\csname cnt@rev.pol.rules.sspro.p1.uniform-fresh@0.25.lam1.score\endcsname{1}
\expandafter\def\csname pct@rev.pol.rules.sspro.p1.uniform-fresh@0.25.lam1.score\endcsname{83.5}
\expandafter\def\csname ppt@rev.pol.rules.sspro.p1.uniform-fresh@0.25.lam1.score\endcsname{$+$83.5}
\expandafter\def\csname val@rev.pol.rules.sspro.p1.uniform-fresh@0.25.m0.5.keep\endcsname{0.797}
\expandafter\def\csname cnt@rev.pol.rules.sspro.p1.uniform-fresh@0.25.m0.5.keep\endcsname{1}
\expandafter\def\csname pct@rev.pol.rules.sspro.p1.uniform-fresh@0.25.m0.5.keep\endcsname{79.7}
\expandafter\def\csname ppt@rev.pol.rules.sspro.p1.uniform-fresh@0.25.m0.5.keep\endcsname{$+$79.7}
\expandafter\def\csname val@rev.pol.rules.sspro.p1.uniform-fresh@0.25.m0.5.req\endcsname{2.372}
\expandafter\def\csname cnt@rev.pol.rules.sspro.p1.uniform-fresh@0.25.m0.5.req\endcsname{2}
\expandafter\def\csname pct@rev.pol.rules.sspro.p1.uniform-fresh@0.25.m0.5.req\endcsname{237.2}
\expandafter\def\csname ppt@rev.pol.rules.sspro.p1.uniform-fresh@0.25.m0.5.req\endcsname{$+$237.2}
\expandafter\def\csname val@rev.pol.rules.sspro.p1.uniform-fresh@0.25.m0.5.score\endcsname{0.835}
\expandafter\def\csname cnt@rev.pol.rules.sspro.p1.uniform-fresh@0.25.m0.5.score\endcsname{1}
\expandafter\def\csname pct@rev.pol.rules.sspro.p1.uniform-fresh@0.25.m0.5.score\endcsname{83.5}
\expandafter\def\csname ppt@rev.pol.rules.sspro.p1.uniform-fresh@0.25.m0.5.score\endcsname{$+$83.5}
\expandafter\def\csname val@rev.pol.rules.sspro.p1.uniform-fresh@0.25.m1.keep\endcsname{0.797}
\expandafter\def\csname cnt@rev.pol.rules.sspro.p1.uniform-fresh@0.25.m1.keep\endcsname{1}
\expandafter\def\csname pct@rev.pol.rules.sspro.p1.uniform-fresh@0.25.m1.keep\endcsname{79.7}
\expandafter\def\csname ppt@rev.pol.rules.sspro.p1.uniform-fresh@0.25.m1.keep\endcsname{$+$79.7}
\expandafter\def\csname val@rev.pol.rules.sspro.p1.uniform-fresh@0.25.m1.req\endcsname{2.372}
\expandafter\def\csname cnt@rev.pol.rules.sspro.p1.uniform-fresh@0.25.m1.req\endcsname{2}
\expandafter\def\csname pct@rev.pol.rules.sspro.p1.uniform-fresh@0.25.m1.req\endcsname{237.2}
\expandafter\def\csname ppt@rev.pol.rules.sspro.p1.uniform-fresh@0.25.m1.req\endcsname{$+$237.2}
\expandafter\def\csname val@rev.pol.rules.sspro.p1.uniform-fresh@0.25.m1.score\endcsname{0.835}
\expandafter\def\csname cnt@rev.pol.rules.sspro.p1.uniform-fresh@0.25.m1.score\endcsname{1}
\expandafter\def\csname pct@rev.pol.rules.sspro.p1.uniform-fresh@0.25.m1.score\endcsname{83.5}
\expandafter\def\csname ppt@rev.pol.rules.sspro.p1.uniform-fresh@0.25.m1.score\endcsname{$+$83.5}
\expandafter\def\csname val@rev.pol.rules.sspro.p1.uniform-fresh@0.25.m2.keep\endcsname{0.797}
\expandafter\def\csname cnt@rev.pol.rules.sspro.p1.uniform-fresh@0.25.m2.keep\endcsname{1}
\expandafter\def\csname pct@rev.pol.rules.sspro.p1.uniform-fresh@0.25.m2.keep\endcsname{79.7}
\expandafter\def\csname ppt@rev.pol.rules.sspro.p1.uniform-fresh@0.25.m2.keep\endcsname{$+$79.7}
\expandafter\def\csname val@rev.pol.rules.sspro.p1.uniform-fresh@0.25.m2.req\endcsname{2.372}
\expandafter\def\csname cnt@rev.pol.rules.sspro.p1.uniform-fresh@0.25.m2.req\endcsname{2}
\expandafter\def\csname pct@rev.pol.rules.sspro.p1.uniform-fresh@0.25.m2.req\endcsname{237.2}
\expandafter\def\csname ppt@rev.pol.rules.sspro.p1.uniform-fresh@0.25.m2.req\endcsname{$+$237.2}
\expandafter\def\csname val@rev.pol.rules.sspro.p1.uniform-fresh@0.25.m2.score\endcsname{0.835}
\expandafter\def\csname cnt@rev.pol.rules.sspro.p1.uniform-fresh@0.25.m2.score\endcsname{1}
\expandafter\def\csname pct@rev.pol.rules.sspro.p1.uniform-fresh@0.25.m2.score\endcsname{83.5}
\expandafter\def\csname ppt@rev.pol.rules.sspro.p1.uniform-fresh@0.25.m2.score\endcsname{$+$83.5}
\expandafter\def\csname val@rev.pol.rules.sspro.p2.full-recert.frozen.keep\endcsname{1.000}
\expandafter\def\csname cnt@rev.pol.rules.sspro.p2.full-recert.frozen.keep\endcsname{1}
\expandafter\def\csname pct@rev.pol.rules.sspro.p2.full-recert.frozen.keep\endcsname{100.0}
\expandafter\def\csname ppt@rev.pol.rules.sspro.p2.full-recert.frozen.keep\endcsname{$+$100.0}
\expandafter\def\csname val@rev.pol.rules.sspro.p2.full-recert.frozen.req\endcsname{2.669}
\expandafter\def\csname cnt@rev.pol.rules.sspro.p2.full-recert.frozen.req\endcsname{3}
\expandafter\def\csname pct@rev.pol.rules.sspro.p2.full-recert.frozen.req\endcsname{266.9}
\expandafter\def\csname ppt@rev.pol.rules.sspro.p2.full-recert.frozen.req\endcsname{$+$266.9}
\expandafter\def\csname val@rev.pol.rules.sspro.p2.full-recert.frozen.score\endcsname{0.906}
\expandafter\def\csname cnt@rev.pol.rules.sspro.p2.full-recert.frozen.score\endcsname{1}
\expandafter\def\csname pct@rev.pol.rules.sspro.p2.full-recert.frozen.score\endcsname{90.6}
\expandafter\def\csname ppt@rev.pol.rules.sspro.p2.full-recert.frozen.score\endcsname{$+$90.6}
\expandafter\def\csname val@rev.pol.rules.sspro.p2.full-recert.lam0.5.keep\endcsname{1.000}
\expandafter\def\csname cnt@rev.pol.rules.sspro.p2.full-recert.lam0.5.keep\endcsname{1}
\expandafter\def\csname pct@rev.pol.rules.sspro.p2.full-recert.lam0.5.keep\endcsname{100.0}
\expandafter\def\csname ppt@rev.pol.rules.sspro.p2.full-recert.lam0.5.keep\endcsname{$+$100.0}
\expandafter\def\csname val@rev.pol.rules.sspro.p2.full-recert.lam0.5.req\endcsname{2.669}
\expandafter\def\csname cnt@rev.pol.rules.sspro.p2.full-recert.lam0.5.req\endcsname{3}
\expandafter\def\csname pct@rev.pol.rules.sspro.p2.full-recert.lam0.5.req\endcsname{266.9}
\expandafter\def\csname ppt@rev.pol.rules.sspro.p2.full-recert.lam0.5.req\endcsname{$+$266.9}
\expandafter\def\csname val@rev.pol.rules.sspro.p2.full-recert.lam0.5.score\endcsname{0.906}
\expandafter\def\csname cnt@rev.pol.rules.sspro.p2.full-recert.lam0.5.score\endcsname{1}
\expandafter\def\csname pct@rev.pol.rules.sspro.p2.full-recert.lam0.5.score\endcsname{90.6}
\expandafter\def\csname ppt@rev.pol.rules.sspro.p2.full-recert.lam0.5.score\endcsname{$+$90.6}
\expandafter\def\csname val@rev.pol.rules.sspro.p2.full-recert.lam1.keep\endcsname{1.000}
\expandafter\def\csname cnt@rev.pol.rules.sspro.p2.full-recert.lam1.keep\endcsname{1}
\expandafter\def\csname pct@rev.pol.rules.sspro.p2.full-recert.lam1.keep\endcsname{100.0}
\expandafter\def\csname ppt@rev.pol.rules.sspro.p2.full-recert.lam1.keep\endcsname{$+$100.0}
\expandafter\def\csname val@rev.pol.rules.sspro.p2.full-recert.lam1.req\endcsname{2.669}
\expandafter\def\csname cnt@rev.pol.rules.sspro.p2.full-recert.lam1.req\endcsname{3}
\expandafter\def\csname pct@rev.pol.rules.sspro.p2.full-recert.lam1.req\endcsname{266.9}
\expandafter\def\csname ppt@rev.pol.rules.sspro.p2.full-recert.lam1.req\endcsname{$+$266.9}
\expandafter\def\csname val@rev.pol.rules.sspro.p2.full-recert.lam1.score\endcsname{0.906}
\expandafter\def\csname cnt@rev.pol.rules.sspro.p2.full-recert.lam1.score\endcsname{1}
\expandafter\def\csname pct@rev.pol.rules.sspro.p2.full-recert.lam1.score\endcsname{90.6}
\expandafter\def\csname ppt@rev.pol.rules.sspro.p2.full-recert.lam1.score\endcsname{$+$90.6}
\expandafter\def\csname val@rev.pol.rules.sspro.p2.full-recert.m0.5.keep\endcsname{1.000}
\expandafter\def\csname cnt@rev.pol.rules.sspro.p2.full-recert.m0.5.keep\endcsname{1}
\expandafter\def\csname pct@rev.pol.rules.sspro.p2.full-recert.m0.5.keep\endcsname{100.0}
\expandafter\def\csname ppt@rev.pol.rules.sspro.p2.full-recert.m0.5.keep\endcsname{$+$100.0}
\expandafter\def\csname val@rev.pol.rules.sspro.p2.full-recert.m0.5.req\endcsname{2.669}
\expandafter\def\csname cnt@rev.pol.rules.sspro.p2.full-recert.m0.5.req\endcsname{3}
\expandafter\def\csname pct@rev.pol.rules.sspro.p2.full-recert.m0.5.req\endcsname{266.9}
\expandafter\def\csname ppt@rev.pol.rules.sspro.p2.full-recert.m0.5.req\endcsname{$+$266.9}
\expandafter\def\csname val@rev.pol.rules.sspro.p2.full-recert.m0.5.score\endcsname{0.906}
\expandafter\def\csname cnt@rev.pol.rules.sspro.p2.full-recert.m0.5.score\endcsname{1}
\expandafter\def\csname pct@rev.pol.rules.sspro.p2.full-recert.m0.5.score\endcsname{90.6}
\expandafter\def\csname ppt@rev.pol.rules.sspro.p2.full-recert.m0.5.score\endcsname{$+$90.6}
\expandafter\def\csname val@rev.pol.rules.sspro.p2.full-recert.m1.keep\endcsname{1.000}
\expandafter\def\csname cnt@rev.pol.rules.sspro.p2.full-recert.m1.keep\endcsname{1}
\expandafter\def\csname pct@rev.pol.rules.sspro.p2.full-recert.m1.keep\endcsname{100.0}
\expandafter\def\csname ppt@rev.pol.rules.sspro.p2.full-recert.m1.keep\endcsname{$+$100.0}
\expandafter\def\csname val@rev.pol.rules.sspro.p2.full-recert.m1.req\endcsname{2.669}
\expandafter\def\csname cnt@rev.pol.rules.sspro.p2.full-recert.m1.req\endcsname{3}
\expandafter\def\csname pct@rev.pol.rules.sspro.p2.full-recert.m1.req\endcsname{266.9}
\expandafter\def\csname ppt@rev.pol.rules.sspro.p2.full-recert.m1.req\endcsname{$+$266.9}
\expandafter\def\csname val@rev.pol.rules.sspro.p2.full-recert.m1.score\endcsname{0.906}
\expandafter\def\csname cnt@rev.pol.rules.sspro.p2.full-recert.m1.score\endcsname{1}
\expandafter\def\csname pct@rev.pol.rules.sspro.p2.full-recert.m1.score\endcsname{90.6}
\expandafter\def\csname ppt@rev.pol.rules.sspro.p2.full-recert.m1.score\endcsname{$+$90.6}
\expandafter\def\csname val@rev.pol.rules.sspro.p2.full-recert.m2.keep\endcsname{1.000}
\expandafter\def\csname cnt@rev.pol.rules.sspro.p2.full-recert.m2.keep\endcsname{1}
\expandafter\def\csname pct@rev.pol.rules.sspro.p2.full-recert.m2.keep\endcsname{100.0}
\expandafter\def\csname ppt@rev.pol.rules.sspro.p2.full-recert.m2.keep\endcsname{$+$100.0}
\expandafter\def\csname val@rev.pol.rules.sspro.p2.full-recert.m2.req\endcsname{2.669}
\expandafter\def\csname cnt@rev.pol.rules.sspro.p2.full-recert.m2.req\endcsname{3}
\expandafter\def\csname pct@rev.pol.rules.sspro.p2.full-recert.m2.req\endcsname{266.9}
\expandafter\def\csname ppt@rev.pol.rules.sspro.p2.full-recert.m2.req\endcsname{$+$266.9}
\expandafter\def\csname val@rev.pol.rules.sspro.p2.full-recert.m2.score\endcsname{0.906}
\expandafter\def\csname cnt@rev.pol.rules.sspro.p2.full-recert.m2.score\endcsname{1}
\expandafter\def\csname pct@rev.pol.rules.sspro.p2.full-recert.m2.score\endcsname{90.6}
\expandafter\def\csname ppt@rev.pol.rules.sspro.p2.full-recert.m2.score\endcsname{$+$90.6}
\expandafter\def\csname val@rev.pol.rules.sspro.p2.generic-ppi-source@0.25.frozen.keep\endcsname{0.898}
\expandafter\def\csname cnt@rev.pol.rules.sspro.p2.generic-ppi-source@0.25.frozen.keep\endcsname{1}
\expandafter\def\csname pct@rev.pol.rules.sspro.p2.generic-ppi-source@0.25.frozen.keep\endcsname{89.8}
\expandafter\def\csname ppt@rev.pol.rules.sspro.p2.generic-ppi-source@0.25.frozen.keep\endcsname{$+$89.8}
\expandafter\def\csname val@rev.pol.rules.sspro.p2.generic-ppi-source@0.25.frozen.req\endcsname{2.447}
\expandafter\def\csname cnt@rev.pol.rules.sspro.p2.generic-ppi-source@0.25.frozen.req\endcsname{2}
\expandafter\def\csname pct@rev.pol.rules.sspro.p2.generic-ppi-source@0.25.frozen.req\endcsname{244.7}
\expandafter\def\csname ppt@rev.pol.rules.sspro.p2.generic-ppi-source@0.25.frozen.req\endcsname{$+$244.7}
\expandafter\def\csname val@rev.pol.rules.sspro.p2.generic-ppi-source@0.25.frozen.score\endcsname{0.891}
\expandafter\def\csname cnt@rev.pol.rules.sspro.p2.generic-ppi-source@0.25.frozen.score\endcsname{1}
\expandafter\def\csname pct@rev.pol.rules.sspro.p2.generic-ppi-source@0.25.frozen.score\endcsname{89.1}
\expandafter\def\csname ppt@rev.pol.rules.sspro.p2.generic-ppi-source@0.25.frozen.score\endcsname{$+$89.1}
\expandafter\def\csname val@rev.pol.rules.sspro.p2.generic-ppi-source@0.25.lam0.5.keep\endcsname{0.898}
\expandafter\def\csname cnt@rev.pol.rules.sspro.p2.generic-ppi-source@0.25.lam0.5.keep\endcsname{1}
\expandafter\def\csname pct@rev.pol.rules.sspro.p2.generic-ppi-source@0.25.lam0.5.keep\endcsname{89.8}
\expandafter\def\csname ppt@rev.pol.rules.sspro.p2.generic-ppi-source@0.25.lam0.5.keep\endcsname{$+$89.8}
\expandafter\def\csname val@rev.pol.rules.sspro.p2.generic-ppi-source@0.25.lam0.5.req\endcsname{2.447}
\expandafter\def\csname cnt@rev.pol.rules.sspro.p2.generic-ppi-source@0.25.lam0.5.req\endcsname{2}
\expandafter\def\csname pct@rev.pol.rules.sspro.p2.generic-ppi-source@0.25.lam0.5.req\endcsname{244.7}
\expandafter\def\csname ppt@rev.pol.rules.sspro.p2.generic-ppi-source@0.25.lam0.5.req\endcsname{$+$244.7}
\expandafter\def\csname val@rev.pol.rules.sspro.p2.generic-ppi-source@0.25.lam0.5.score\endcsname{0.891}
\expandafter\def\csname cnt@rev.pol.rules.sspro.p2.generic-ppi-source@0.25.lam0.5.score\endcsname{1}
\expandafter\def\csname pct@rev.pol.rules.sspro.p2.generic-ppi-source@0.25.lam0.5.score\endcsname{89.1}
\expandafter\def\csname ppt@rev.pol.rules.sspro.p2.generic-ppi-source@0.25.lam0.5.score\endcsname{$+$89.1}
\expandafter\def\csname val@rev.pol.rules.sspro.p2.generic-ppi-source@0.25.lam1.keep\endcsname{0.898}
\expandafter\def\csname cnt@rev.pol.rules.sspro.p2.generic-ppi-source@0.25.lam1.keep\endcsname{1}
\expandafter\def\csname pct@rev.pol.rules.sspro.p2.generic-ppi-source@0.25.lam1.keep\endcsname{89.8}
\expandafter\def\csname ppt@rev.pol.rules.sspro.p2.generic-ppi-source@0.25.lam1.keep\endcsname{$+$89.8}
\expandafter\def\csname val@rev.pol.rules.sspro.p2.generic-ppi-source@0.25.lam1.req\endcsname{2.447}
\expandafter\def\csname cnt@rev.pol.rules.sspro.p2.generic-ppi-source@0.25.lam1.req\endcsname{2}
\expandafter\def\csname pct@rev.pol.rules.sspro.p2.generic-ppi-source@0.25.lam1.req\endcsname{244.7}
\expandafter\def\csname ppt@rev.pol.rules.sspro.p2.generic-ppi-source@0.25.lam1.req\endcsname{$+$244.7}
\expandafter\def\csname val@rev.pol.rules.sspro.p2.generic-ppi-source@0.25.lam1.score\endcsname{0.891}
\expandafter\def\csname cnt@rev.pol.rules.sspro.p2.generic-ppi-source@0.25.lam1.score\endcsname{1}
\expandafter\def\csname pct@rev.pol.rules.sspro.p2.generic-ppi-source@0.25.lam1.score\endcsname{89.1}
\expandafter\def\csname ppt@rev.pol.rules.sspro.p2.generic-ppi-source@0.25.lam1.score\endcsname{$+$89.1}
\expandafter\def\csname val@rev.pol.rules.sspro.p2.generic-ppi-source@0.25.m0.5.keep\endcsname{0.898}
\expandafter\def\csname cnt@rev.pol.rules.sspro.p2.generic-ppi-source@0.25.m0.5.keep\endcsname{1}
\expandafter\def\csname pct@rev.pol.rules.sspro.p2.generic-ppi-source@0.25.m0.5.keep\endcsname{89.8}
\expandafter\def\csname ppt@rev.pol.rules.sspro.p2.generic-ppi-source@0.25.m0.5.keep\endcsname{$+$89.8}
\expandafter\def\csname val@rev.pol.rules.sspro.p2.generic-ppi-source@0.25.m0.5.req\endcsname{2.447}
\expandafter\def\csname cnt@rev.pol.rules.sspro.p2.generic-ppi-source@0.25.m0.5.req\endcsname{2}
\expandafter\def\csname pct@rev.pol.rules.sspro.p2.generic-ppi-source@0.25.m0.5.req\endcsname{244.7}
\expandafter\def\csname ppt@rev.pol.rules.sspro.p2.generic-ppi-source@0.25.m0.5.req\endcsname{$+$244.7}
\expandafter\def\csname val@rev.pol.rules.sspro.p2.generic-ppi-source@0.25.m0.5.score\endcsname{0.891}
\expandafter\def\csname cnt@rev.pol.rules.sspro.p2.generic-ppi-source@0.25.m0.5.score\endcsname{1}
\expandafter\def\csname pct@rev.pol.rules.sspro.p2.generic-ppi-source@0.25.m0.5.score\endcsname{89.1}
\expandafter\def\csname ppt@rev.pol.rules.sspro.p2.generic-ppi-source@0.25.m0.5.score\endcsname{$+$89.1}
\expandafter\def\csname val@rev.pol.rules.sspro.p2.generic-ppi-source@0.25.m1.keep\endcsname{0.898}
\expandafter\def\csname cnt@rev.pol.rules.sspro.p2.generic-ppi-source@0.25.m1.keep\endcsname{1}
\expandafter\def\csname pct@rev.pol.rules.sspro.p2.generic-ppi-source@0.25.m1.keep\endcsname{89.8}
\expandafter\def\csname ppt@rev.pol.rules.sspro.p2.generic-ppi-source@0.25.m1.keep\endcsname{$+$89.8}
\expandafter\def\csname val@rev.pol.rules.sspro.p2.generic-ppi-source@0.25.m1.req\endcsname{2.447}
\expandafter\def\csname cnt@rev.pol.rules.sspro.p2.generic-ppi-source@0.25.m1.req\endcsname{2}
\expandafter\def\csname pct@rev.pol.rules.sspro.p2.generic-ppi-source@0.25.m1.req\endcsname{244.7}
\expandafter\def\csname ppt@rev.pol.rules.sspro.p2.generic-ppi-source@0.25.m1.req\endcsname{$+$244.7}
\expandafter\def\csname val@rev.pol.rules.sspro.p2.generic-ppi-source@0.25.m1.score\endcsname{0.891}
\expandafter\def\csname cnt@rev.pol.rules.sspro.p2.generic-ppi-source@0.25.m1.score\endcsname{1}
\expandafter\def\csname pct@rev.pol.rules.sspro.p2.generic-ppi-source@0.25.m1.score\endcsname{89.1}
\expandafter\def\csname ppt@rev.pol.rules.sspro.p2.generic-ppi-source@0.25.m1.score\endcsname{$+$89.1}
\expandafter\def\csname val@rev.pol.rules.sspro.p2.generic-ppi-source@0.25.m2.keep\endcsname{0.898}
\expandafter\def\csname cnt@rev.pol.rules.sspro.p2.generic-ppi-source@0.25.m2.keep\endcsname{1}
\expandafter\def\csname pct@rev.pol.rules.sspro.p2.generic-ppi-source@0.25.m2.keep\endcsname{89.8}
\expandafter\def\csname ppt@rev.pol.rules.sspro.p2.generic-ppi-source@0.25.m2.keep\endcsname{$+$89.8}
\expandafter\def\csname val@rev.pol.rules.sspro.p2.generic-ppi-source@0.25.m2.req\endcsname{2.447}
\expandafter\def\csname cnt@rev.pol.rules.sspro.p2.generic-ppi-source@0.25.m2.req\endcsname{2}
\expandafter\def\csname pct@rev.pol.rules.sspro.p2.generic-ppi-source@0.25.m2.req\endcsname{244.7}
\expandafter\def\csname ppt@rev.pol.rules.sspro.p2.generic-ppi-source@0.25.m2.req\endcsname{$+$244.7}
\expandafter\def\csname val@rev.pol.rules.sspro.p2.generic-ppi-source@0.25.m2.score\endcsname{0.891}
\expandafter\def\csname cnt@rev.pol.rules.sspro.p2.generic-ppi-source@0.25.m2.score\endcsname{1}
\expandafter\def\csname pct@rev.pol.rules.sspro.p2.generic-ppi-source@0.25.m2.score\endcsname{89.1}
\expandafter\def\csname ppt@rev.pol.rules.sspro.p2.generic-ppi-source@0.25.m2.score\endcsname{$+$89.1}
\expandafter\def\csname val@rev.pol.rules.sspro.p2.keep-all.keep\endcsname{1.000}
\expandafter\def\csname cnt@rev.pol.rules.sspro.p2.keep-all.keep\endcsname{1}
\expandafter\def\csname pct@rev.pol.rules.sspro.p2.keep-all.keep\endcsname{100.0}
\expandafter\def\csname ppt@rev.pol.rules.sspro.p2.keep-all.keep\endcsname{$+$100.0}
\expandafter\def\csname val@rev.pol.rules.sspro.p2.keep-all.req\endcsname{2.669}
\expandafter\def\csname cnt@rev.pol.rules.sspro.p2.keep-all.req\endcsname{3}
\expandafter\def\csname pct@rev.pol.rules.sspro.p2.keep-all.req\endcsname{266.9}
\expandafter\def\csname ppt@rev.pol.rules.sspro.p2.keep-all.req\endcsname{$+$266.9}
\expandafter\def\csname val@rev.pol.rules.sspro.p2.keep-all.score\endcsname{0.906}
\expandafter\def\csname cnt@rev.pol.rules.sspro.p2.keep-all.score\endcsname{1}
\expandafter\def\csname pct@rev.pol.rules.sspro.p2.keep-all.score\endcsname{90.6}
\expandafter\def\csname ppt@rev.pol.rules.sspro.p2.keep-all.score\endcsname{$+$90.6}
\expandafter\def\csname val@rev.pol.rules.sspro.p2.old-decision.frozen.keep\endcsname{1.000}
\expandafter\def\csname cnt@rev.pol.rules.sspro.p2.old-decision.frozen.keep\endcsname{1}
\expandafter\def\csname pct@rev.pol.rules.sspro.p2.old-decision.frozen.keep\endcsname{100.0}
\expandafter\def\csname ppt@rev.pol.rules.sspro.p2.old-decision.frozen.keep\endcsname{$+$100.0}
\expandafter\def\csname val@rev.pol.rules.sspro.p2.old-decision.frozen.req\endcsname{2.669}
\expandafter\def\csname cnt@rev.pol.rules.sspro.p2.old-decision.frozen.req\endcsname{3}
\expandafter\def\csname pct@rev.pol.rules.sspro.p2.old-decision.frozen.req\endcsname{266.9}
\expandafter\def\csname ppt@rev.pol.rules.sspro.p2.old-decision.frozen.req\endcsname{$+$266.9}
\expandafter\def\csname val@rev.pol.rules.sspro.p2.old-decision.frozen.score\endcsname{0.906}
\expandafter\def\csname cnt@rev.pol.rules.sspro.p2.old-decision.frozen.score\endcsname{1}
\expandafter\def\csname pct@rev.pol.rules.sspro.p2.old-decision.frozen.score\endcsname{90.6}
\expandafter\def\csname ppt@rev.pol.rules.sspro.p2.old-decision.frozen.score\endcsname{$+$90.6}
\expandafter\def\csname val@rev.pol.rules.sspro.p2.old-decision.lam0.5.keep\endcsname{1.000}
\expandafter\def\csname cnt@rev.pol.rules.sspro.p2.old-decision.lam0.5.keep\endcsname{1}
\expandafter\def\csname pct@rev.pol.rules.sspro.p2.old-decision.lam0.5.keep\endcsname{100.0}
\expandafter\def\csname ppt@rev.pol.rules.sspro.p2.old-decision.lam0.5.keep\endcsname{$+$100.0}
\expandafter\def\csname val@rev.pol.rules.sspro.p2.old-decision.lam0.5.req\endcsname{2.669}
\expandafter\def\csname cnt@rev.pol.rules.sspro.p2.old-decision.lam0.5.req\endcsname{3}
\expandafter\def\csname pct@rev.pol.rules.sspro.p2.old-decision.lam0.5.req\endcsname{266.9}
\expandafter\def\csname ppt@rev.pol.rules.sspro.p2.old-decision.lam0.5.req\endcsname{$+$266.9}
\expandafter\def\csname val@rev.pol.rules.sspro.p2.old-decision.lam0.5.score\endcsname{0.906}
\expandafter\def\csname cnt@rev.pol.rules.sspro.p2.old-decision.lam0.5.score\endcsname{1}
\expandafter\def\csname pct@rev.pol.rules.sspro.p2.old-decision.lam0.5.score\endcsname{90.6}
\expandafter\def\csname ppt@rev.pol.rules.sspro.p2.old-decision.lam0.5.score\endcsname{$+$90.6}
\expandafter\def\csname val@rev.pol.rules.sspro.p2.old-decision.lam1.keep\endcsname{1.000}
\expandafter\def\csname cnt@rev.pol.rules.sspro.p2.old-decision.lam1.keep\endcsname{1}
\expandafter\def\csname pct@rev.pol.rules.sspro.p2.old-decision.lam1.keep\endcsname{100.0}
\expandafter\def\csname ppt@rev.pol.rules.sspro.p2.old-decision.lam1.keep\endcsname{$+$100.0}
\expandafter\def\csname val@rev.pol.rules.sspro.p2.old-decision.lam1.req\endcsname{2.669}
\expandafter\def\csname cnt@rev.pol.rules.sspro.p2.old-decision.lam1.req\endcsname{3}
\expandafter\def\csname pct@rev.pol.rules.sspro.p2.old-decision.lam1.req\endcsname{266.9}
\expandafter\def\csname ppt@rev.pol.rules.sspro.p2.old-decision.lam1.req\endcsname{$+$266.9}
\expandafter\def\csname val@rev.pol.rules.sspro.p2.old-decision.lam1.score\endcsname{0.906}
\expandafter\def\csname cnt@rev.pol.rules.sspro.p2.old-decision.lam1.score\endcsname{1}
\expandafter\def\csname pct@rev.pol.rules.sspro.p2.old-decision.lam1.score\endcsname{90.6}
\expandafter\def\csname ppt@rev.pol.rules.sspro.p2.old-decision.lam1.score\endcsname{$+$90.6}
\expandafter\def\csname val@rev.pol.rules.sspro.p2.old-decision.m0.5.keep\endcsname{1.000}
\expandafter\def\csname cnt@rev.pol.rules.sspro.p2.old-decision.m0.5.keep\endcsname{1}
\expandafter\def\csname pct@rev.pol.rules.sspro.p2.old-decision.m0.5.keep\endcsname{100.0}
\expandafter\def\csname ppt@rev.pol.rules.sspro.p2.old-decision.m0.5.keep\endcsname{$+$100.0}
\expandafter\def\csname val@rev.pol.rules.sspro.p2.old-decision.m0.5.req\endcsname{2.669}
\expandafter\def\csname cnt@rev.pol.rules.sspro.p2.old-decision.m0.5.req\endcsname{3}
\expandafter\def\csname pct@rev.pol.rules.sspro.p2.old-decision.m0.5.req\endcsname{266.9}
\expandafter\def\csname ppt@rev.pol.rules.sspro.p2.old-decision.m0.5.req\endcsname{$+$266.9}
\expandafter\def\csname val@rev.pol.rules.sspro.p2.old-decision.m0.5.score\endcsname{0.906}
\expandafter\def\csname cnt@rev.pol.rules.sspro.p2.old-decision.m0.5.score\endcsname{1}
\expandafter\def\csname pct@rev.pol.rules.sspro.p2.old-decision.m0.5.score\endcsname{90.6}
\expandafter\def\csname ppt@rev.pol.rules.sspro.p2.old-decision.m0.5.score\endcsname{$+$90.6}
\expandafter\def\csname val@rev.pol.rules.sspro.p2.old-decision.m1.keep\endcsname{1.000}
\expandafter\def\csname cnt@rev.pol.rules.sspro.p2.old-decision.m1.keep\endcsname{1}
\expandafter\def\csname pct@rev.pol.rules.sspro.p2.old-decision.m1.keep\endcsname{100.0}
\expandafter\def\csname ppt@rev.pol.rules.sspro.p2.old-decision.m1.keep\endcsname{$+$100.0}
\expandafter\def\csname val@rev.pol.rules.sspro.p2.old-decision.m1.req\endcsname{2.669}
\expandafter\def\csname cnt@rev.pol.rules.sspro.p2.old-decision.m1.req\endcsname{3}
\expandafter\def\csname pct@rev.pol.rules.sspro.p2.old-decision.m1.req\endcsname{266.9}
\expandafter\def\csname ppt@rev.pol.rules.sspro.p2.old-decision.m1.req\endcsname{$+$266.9}
\expandafter\def\csname val@rev.pol.rules.sspro.p2.old-decision.m1.score\endcsname{0.906}
\expandafter\def\csname cnt@rev.pol.rules.sspro.p2.old-decision.m1.score\endcsname{1}
\expandafter\def\csname pct@rev.pol.rules.sspro.p2.old-decision.m1.score\endcsname{90.6}
\expandafter\def\csname ppt@rev.pol.rules.sspro.p2.old-decision.m1.score\endcsname{$+$90.6}
\expandafter\def\csname val@rev.pol.rules.sspro.p2.old-decision.m2.keep\endcsname{1.000}
\expandafter\def\csname cnt@rev.pol.rules.sspro.p2.old-decision.m2.keep\endcsname{1}
\expandafter\def\csname pct@rev.pol.rules.sspro.p2.old-decision.m2.keep\endcsname{100.0}
\expandafter\def\csname ppt@rev.pol.rules.sspro.p2.old-decision.m2.keep\endcsname{$+$100.0}
\expandafter\def\csname val@rev.pol.rules.sspro.p2.old-decision.m2.req\endcsname{2.669}
\expandafter\def\csname cnt@rev.pol.rules.sspro.p2.old-decision.m2.req\endcsname{3}
\expandafter\def\csname pct@rev.pol.rules.sspro.p2.old-decision.m2.req\endcsname{266.9}
\expandafter\def\csname ppt@rev.pol.rules.sspro.p2.old-decision.m2.req\endcsname{$+$266.9}
\expandafter\def\csname val@rev.pol.rules.sspro.p2.old-decision.m2.score\endcsname{0.906}
\expandafter\def\csname cnt@rev.pol.rules.sspro.p2.old-decision.m2.score\endcsname{1}
\expandafter\def\csname pct@rev.pol.rules.sspro.p2.old-decision.m2.score\endcsname{90.6}
\expandafter\def\csname ppt@rev.pol.rules.sspro.p2.old-decision.m2.score\endcsname{$+$90.6}
\expandafter\def\csname val@rev.pol.rules.sspro.p2.oracle-on-T.keep\endcsname{1.000}
\expandafter\def\csname cnt@rev.pol.rules.sspro.p2.oracle-on-T.keep\endcsname{1}
\expandafter\def\csname pct@rev.pol.rules.sspro.p2.oracle-on-T.keep\endcsname{100.0}
\expandafter\def\csname ppt@rev.pol.rules.sspro.p2.oracle-on-T.keep\endcsname{$+$100.0}
\expandafter\def\csname val@rev.pol.rules.sspro.p2.oracle-on-T.req\endcsname{2.669}
\expandafter\def\csname cnt@rev.pol.rules.sspro.p2.oracle-on-T.req\endcsname{3}
\expandafter\def\csname pct@rev.pol.rules.sspro.p2.oracle-on-T.req\endcsname{266.9}
\expandafter\def\csname ppt@rev.pol.rules.sspro.p2.oracle-on-T.req\endcsname{$+$266.9}
\expandafter\def\csname val@rev.pol.rules.sspro.p2.oracle-on-T.score\endcsname{0.906}
\expandafter\def\csname cnt@rev.pol.rules.sspro.p2.oracle-on-T.score\endcsname{1}
\expandafter\def\csname pct@rev.pol.rules.sspro.p2.oracle-on-T.score\endcsname{90.6}
\expandafter\def\csname ppt@rev.pol.rules.sspro.p2.oracle-on-T.score\endcsname{$+$90.6}
\expandafter\def\csname val@rev.pol.rules.sspro.p2.r-s.center-d1\endcsname{2.000}
\expandafter\def\csname cnt@rev.pol.rules.sspro.p2.r-s.center-d1\endcsname{2}
\expandafter\def\csname pct@rev.pol.rules.sspro.p2.r-s.center-d1\endcsname{200.0}
\expandafter\def\csname ppt@rev.pol.rules.sspro.p2.r-s.center-d1\endcsname{$+$200.0}
\expandafter\def\csname val@rev.pol.rules.sspro.p2.r-s.context-d1\endcsname{2.000}
\expandafter\def\csname cnt@rev.pol.rules.sspro.p2.r-s.context-d1\endcsname{2}
\expandafter\def\csname pct@rev.pol.rules.sspro.p2.r-s.context-d1\endcsname{200.0}
\expandafter\def\csname ppt@rev.pol.rules.sspro.p2.r-s.context-d1\endcsname{$+$200.0}
\expandafter\def\csname val@rev.pol.rules.sspro.p2.r-s.dual-d2\endcsname{3.000}
\expandafter\def\csname cnt@rev.pol.rules.sspro.p2.r-s.dual-d2\endcsname{3}
\expandafter\def\csname pct@rev.pol.rules.sspro.p2.r-s.dual-d2\endcsname{300.0}
\expandafter\def\csname ppt@rev.pol.rules.sspro.p2.r-s.dual-d2\endcsname{$+$300.0}
\expandafter\def\csname val@rev.pol.rules.sspro.p2.r-s.grid2-d3\endcsname{3.917}
\expandafter\def\csname cnt@rev.pol.rules.sspro.p2.r-s.grid2-d3\endcsname{4}
\expandafter\def\csname pct@rev.pol.rules.sspro.p2.r-s.grid2-d3\endcsname{391.7}
\expandafter\def\csname ppt@rev.pol.rules.sspro.p2.r-s.grid2-d3\endcsname{$+$391.7}
\expandafter\def\csname val@rev.pol.rules.sspro.p2.rcr-stratified@0.25.frozen.keep\endcsname{0.897}
\expandafter\def\csname cnt@rev.pol.rules.sspro.p2.rcr-stratified@0.25.frozen.keep\endcsname{1}
\expandafter\def\csname pct@rev.pol.rules.sspro.p2.rcr-stratified@0.25.frozen.keep\endcsname{89.7}
\expandafter\def\csname ppt@rev.pol.rules.sspro.p2.rcr-stratified@0.25.frozen.keep\endcsname{$+$89.7}
\expandafter\def\csname val@rev.pol.rules.sspro.p2.rcr-stratified@0.25.frozen.req\endcsname{2.438}
\expandafter\def\csname cnt@rev.pol.rules.sspro.p2.rcr-stratified@0.25.frozen.req\endcsname{2}
\expandafter\def\csname pct@rev.pol.rules.sspro.p2.rcr-stratified@0.25.frozen.req\endcsname{243.8}
\expandafter\def\csname ppt@rev.pol.rules.sspro.p2.rcr-stratified@0.25.frozen.req\endcsname{$+$243.8}
\expandafter\def\csname val@rev.pol.rules.sspro.p2.rcr-stratified@0.25.frozen.score\endcsname{0.891}
\expandafter\def\csname cnt@rev.pol.rules.sspro.p2.rcr-stratified@0.25.frozen.score\endcsname{1}
\expandafter\def\csname pct@rev.pol.rules.sspro.p2.rcr-stratified@0.25.frozen.score\endcsname{89.1}
\expandafter\def\csname ppt@rev.pol.rules.sspro.p2.rcr-stratified@0.25.frozen.score\endcsname{$+$89.1}
\expandafter\def\csname val@rev.pol.rules.sspro.p2.rcr-stratified@0.25.lam0.5.keep\endcsname{0.856}
\expandafter\def\csname cnt@rev.pol.rules.sspro.p2.rcr-stratified@0.25.lam0.5.keep\endcsname{1}
\expandafter\def\csname pct@rev.pol.rules.sspro.p2.rcr-stratified@0.25.lam0.5.keep\endcsname{85.6}
\expandafter\def\csname ppt@rev.pol.rules.sspro.p2.rcr-stratified@0.25.lam0.5.keep\endcsname{$+$85.6}
\expandafter\def\csname val@rev.pol.rules.sspro.p2.rcr-stratified@0.25.lam0.5.req\endcsname{2.396}
\expandafter\def\csname cnt@rev.pol.rules.sspro.p2.rcr-stratified@0.25.lam0.5.req\endcsname{2}
\expandafter\def\csname pct@rev.pol.rules.sspro.p2.rcr-stratified@0.25.lam0.5.req\endcsname{239.6}
\expandafter\def\csname ppt@rev.pol.rules.sspro.p2.rcr-stratified@0.25.lam0.5.req\endcsname{$+$239.6}
\expandafter\def\csname val@rev.pol.rules.sspro.p2.rcr-stratified@0.25.lam0.5.score\endcsname{0.885}
\expandafter\def\csname cnt@rev.pol.rules.sspro.p2.rcr-stratified@0.25.lam0.5.score\endcsname{1}
\expandafter\def\csname pct@rev.pol.rules.sspro.p2.rcr-stratified@0.25.lam0.5.score\endcsname{88.5}
\expandafter\def\csname ppt@rev.pol.rules.sspro.p2.rcr-stratified@0.25.lam0.5.score\endcsname{$+$88.5}
\expandafter\def\csname val@rev.pol.rules.sspro.p2.rcr-stratified@0.25.lam1.keep\endcsname{0.829}
\expandafter\def\csname cnt@rev.pol.rules.sspro.p2.rcr-stratified@0.25.lam1.keep\endcsname{1}
\expandafter\def\csname pct@rev.pol.rules.sspro.p2.rcr-stratified@0.25.lam1.keep\endcsname{82.9}
\expandafter\def\csname ppt@rev.pol.rules.sspro.p2.rcr-stratified@0.25.lam1.keep\endcsname{$+$82.9}
\expandafter\def\csname val@rev.pol.rules.sspro.p2.rcr-stratified@0.25.lam1.req\endcsname{2.343}
\expandafter\def\csname cnt@rev.pol.rules.sspro.p2.rcr-stratified@0.25.lam1.req\endcsname{2}
\expandafter\def\csname pct@rev.pol.rules.sspro.p2.rcr-stratified@0.25.lam1.req\endcsname{234.3}
\expandafter\def\csname ppt@rev.pol.rules.sspro.p2.rcr-stratified@0.25.lam1.req\endcsname{$+$234.3}
\expandafter\def\csname val@rev.pol.rules.sspro.p2.rcr-stratified@0.25.lam1.score\endcsname{0.881}
\expandafter\def\csname cnt@rev.pol.rules.sspro.p2.rcr-stratified@0.25.lam1.score\endcsname{1}
\expandafter\def\csname pct@rev.pol.rules.sspro.p2.rcr-stratified@0.25.lam1.score\endcsname{88.1}
\expandafter\def\csname ppt@rev.pol.rules.sspro.p2.rcr-stratified@0.25.lam1.score\endcsname{$+$88.1}
\expandafter\def\csname val@rev.pol.rules.sspro.p2.rcr-stratified@0.25.m0.5.keep\endcsname{0.856}
\expandafter\def\csname cnt@rev.pol.rules.sspro.p2.rcr-stratified@0.25.m0.5.keep\endcsname{1}
\expandafter\def\csname pct@rev.pol.rules.sspro.p2.rcr-stratified@0.25.m0.5.keep\endcsname{85.6}
\expandafter\def\csname ppt@rev.pol.rules.sspro.p2.rcr-stratified@0.25.m0.5.keep\endcsname{$+$85.6}
\expandafter\def\csname val@rev.pol.rules.sspro.p2.rcr-stratified@0.25.m0.5.req\endcsname{2.397}
\expandafter\def\csname cnt@rev.pol.rules.sspro.p2.rcr-stratified@0.25.m0.5.req\endcsname{2}
\expandafter\def\csname pct@rev.pol.rules.sspro.p2.rcr-stratified@0.25.m0.5.req\endcsname{239.7}
\expandafter\def\csname ppt@rev.pol.rules.sspro.p2.rcr-stratified@0.25.m0.5.req\endcsname{$+$239.7}
\expandafter\def\csname val@rev.pol.rules.sspro.p2.rcr-stratified@0.25.m0.5.score\endcsname{0.885}
\expandafter\def\csname cnt@rev.pol.rules.sspro.p2.rcr-stratified@0.25.m0.5.score\endcsname{1}
\expandafter\def\csname pct@rev.pol.rules.sspro.p2.rcr-stratified@0.25.m0.5.score\endcsname{88.5}
\expandafter\def\csname ppt@rev.pol.rules.sspro.p2.rcr-stratified@0.25.m0.5.score\endcsname{$+$88.5}
\expandafter\def\csname val@rev.pol.rules.sspro.p2.rcr-stratified@0.25.m1.keep\endcsname{0.856}
\expandafter\def\csname cnt@rev.pol.rules.sspro.p2.rcr-stratified@0.25.m1.keep\endcsname{1}
\expandafter\def\csname pct@rev.pol.rules.sspro.p2.rcr-stratified@0.25.m1.keep\endcsname{85.6}
\expandafter\def\csname ppt@rev.pol.rules.sspro.p2.rcr-stratified@0.25.m1.keep\endcsname{$+$85.6}
\expandafter\def\csname val@rev.pol.rules.sspro.p2.rcr-stratified@0.25.m1.req\endcsname{2.396}
\expandafter\def\csname cnt@rev.pol.rules.sspro.p2.rcr-stratified@0.25.m1.req\endcsname{2}
\expandafter\def\csname pct@rev.pol.rules.sspro.p2.rcr-stratified@0.25.m1.req\endcsname{239.6}
\expandafter\def\csname ppt@rev.pol.rules.sspro.p2.rcr-stratified@0.25.m1.req\endcsname{$+$239.6}
\expandafter\def\csname val@rev.pol.rules.sspro.p2.rcr-stratified@0.25.m1.score\endcsname{0.885}
\expandafter\def\csname cnt@rev.pol.rules.sspro.p2.rcr-stratified@0.25.m1.score\endcsname{1}
\expandafter\def\csname pct@rev.pol.rules.sspro.p2.rcr-stratified@0.25.m1.score\endcsname{88.5}
\expandafter\def\csname ppt@rev.pol.rules.sspro.p2.rcr-stratified@0.25.m1.score\endcsname{$+$88.5}
\expandafter\def\csname val@rev.pol.rules.sspro.p2.rcr-stratified@0.25.m2.keep\endcsname{0.829}
\expandafter\def\csname cnt@rev.pol.rules.sspro.p2.rcr-stratified@0.25.m2.keep\endcsname{1}
\expandafter\def\csname pct@rev.pol.rules.sspro.p2.rcr-stratified@0.25.m2.keep\endcsname{82.9}
\expandafter\def\csname ppt@rev.pol.rules.sspro.p2.rcr-stratified@0.25.m2.keep\endcsname{$+$82.9}
\expandafter\def\csname val@rev.pol.rules.sspro.p2.rcr-stratified@0.25.m2.req\endcsname{2.343}
\expandafter\def\csname cnt@rev.pol.rules.sspro.p2.rcr-stratified@0.25.m2.req\endcsname{2}
\expandafter\def\csname pct@rev.pol.rules.sspro.p2.rcr-stratified@0.25.m2.req\endcsname{234.3}
\expandafter\def\csname ppt@rev.pol.rules.sspro.p2.rcr-stratified@0.25.m2.req\endcsname{$+$234.3}
\expandafter\def\csname val@rev.pol.rules.sspro.p2.rcr-stratified@0.25.m2.score\endcsname{0.881}
\expandafter\def\csname cnt@rev.pol.rules.sspro.p2.rcr-stratified@0.25.m2.score\endcsname{1}
\expandafter\def\csname pct@rev.pol.rules.sspro.p2.rcr-stratified@0.25.m2.score\endcsname{88.1}
\expandafter\def\csname ppt@rev.pol.rules.sspro.p2.rcr-stratified@0.25.m2.score\endcsname{$+$88.1}
\expandafter\def\csname val@rev.pol.rules.sspro.p2.replay-only.frozen.keep\endcsname{1.000}
\expandafter\def\csname cnt@rev.pol.rules.sspro.p2.replay-only.frozen.keep\endcsname{1}
\expandafter\def\csname pct@rev.pol.rules.sspro.p2.replay-only.frozen.keep\endcsname{100.0}
\expandafter\def\csname ppt@rev.pol.rules.sspro.p2.replay-only.frozen.keep\endcsname{$+$100.0}
\expandafter\def\csname val@rev.pol.rules.sspro.p2.replay-only.frozen.req\endcsname{2.669}
\expandafter\def\csname cnt@rev.pol.rules.sspro.p2.replay-only.frozen.req\endcsname{3}
\expandafter\def\csname pct@rev.pol.rules.sspro.p2.replay-only.frozen.req\endcsname{266.9}
\expandafter\def\csname ppt@rev.pol.rules.sspro.p2.replay-only.frozen.req\endcsname{$+$266.9}
\expandafter\def\csname val@rev.pol.rules.sspro.p2.replay-only.frozen.score\endcsname{0.906}
\expandafter\def\csname cnt@rev.pol.rules.sspro.p2.replay-only.frozen.score\endcsname{1}
\expandafter\def\csname pct@rev.pol.rules.sspro.p2.replay-only.frozen.score\endcsname{90.6}
\expandafter\def\csname ppt@rev.pol.rules.sspro.p2.replay-only.frozen.score\endcsname{$+$90.6}
\expandafter\def\csname val@rev.pol.rules.sspro.p2.replay-only.lam0.5.keep\endcsname{1.000}
\expandafter\def\csname cnt@rev.pol.rules.sspro.p2.replay-only.lam0.5.keep\endcsname{1}
\expandafter\def\csname pct@rev.pol.rules.sspro.p2.replay-only.lam0.5.keep\endcsname{100.0}
\expandafter\def\csname ppt@rev.pol.rules.sspro.p2.replay-only.lam0.5.keep\endcsname{$+$100.0}
\expandafter\def\csname val@rev.pol.rules.sspro.p2.replay-only.lam0.5.req\endcsname{2.669}
\expandafter\def\csname cnt@rev.pol.rules.sspro.p2.replay-only.lam0.5.req\endcsname{3}
\expandafter\def\csname pct@rev.pol.rules.sspro.p2.replay-only.lam0.5.req\endcsname{266.9}
\expandafter\def\csname ppt@rev.pol.rules.sspro.p2.replay-only.lam0.5.req\endcsname{$+$266.9}
\expandafter\def\csname val@rev.pol.rules.sspro.p2.replay-only.lam0.5.score\endcsname{0.906}
\expandafter\def\csname cnt@rev.pol.rules.sspro.p2.replay-only.lam0.5.score\endcsname{1}
\expandafter\def\csname pct@rev.pol.rules.sspro.p2.replay-only.lam0.5.score\endcsname{90.6}
\expandafter\def\csname ppt@rev.pol.rules.sspro.p2.replay-only.lam0.5.score\endcsname{$+$90.6}
\expandafter\def\csname val@rev.pol.rules.sspro.p2.replay-only.lam1.keep\endcsname{1.000}
\expandafter\def\csname cnt@rev.pol.rules.sspro.p2.replay-only.lam1.keep\endcsname{1}
\expandafter\def\csname pct@rev.pol.rules.sspro.p2.replay-only.lam1.keep\endcsname{100.0}
\expandafter\def\csname ppt@rev.pol.rules.sspro.p2.replay-only.lam1.keep\endcsname{$+$100.0}
\expandafter\def\csname val@rev.pol.rules.sspro.p2.replay-only.lam1.req\endcsname{2.669}
\expandafter\def\csname cnt@rev.pol.rules.sspro.p2.replay-only.lam1.req\endcsname{3}
\expandafter\def\csname pct@rev.pol.rules.sspro.p2.replay-only.lam1.req\endcsname{266.9}
\expandafter\def\csname ppt@rev.pol.rules.sspro.p2.replay-only.lam1.req\endcsname{$+$266.9}
\expandafter\def\csname val@rev.pol.rules.sspro.p2.replay-only.lam1.score\endcsname{0.906}
\expandafter\def\csname cnt@rev.pol.rules.sspro.p2.replay-only.lam1.score\endcsname{1}
\expandafter\def\csname pct@rev.pol.rules.sspro.p2.replay-only.lam1.score\endcsname{90.6}
\expandafter\def\csname ppt@rev.pol.rules.sspro.p2.replay-only.lam1.score\endcsname{$+$90.6}
\expandafter\def\csname val@rev.pol.rules.sspro.p2.replay-only.m0.5.keep\endcsname{1.000}
\expandafter\def\csname cnt@rev.pol.rules.sspro.p2.replay-only.m0.5.keep\endcsname{1}
\expandafter\def\csname pct@rev.pol.rules.sspro.p2.replay-only.m0.5.keep\endcsname{100.0}
\expandafter\def\csname ppt@rev.pol.rules.sspro.p2.replay-only.m0.5.keep\endcsname{$+$100.0}
\expandafter\def\csname val@rev.pol.rules.sspro.p2.replay-only.m0.5.req\endcsname{2.669}
\expandafter\def\csname cnt@rev.pol.rules.sspro.p2.replay-only.m0.5.req\endcsname{3}
\expandafter\def\csname pct@rev.pol.rules.sspro.p2.replay-only.m0.5.req\endcsname{266.9}
\expandafter\def\csname ppt@rev.pol.rules.sspro.p2.replay-only.m0.5.req\endcsname{$+$266.9}
\expandafter\def\csname val@rev.pol.rules.sspro.p2.replay-only.m0.5.score\endcsname{0.906}
\expandafter\def\csname cnt@rev.pol.rules.sspro.p2.replay-only.m0.5.score\endcsname{1}
\expandafter\def\csname pct@rev.pol.rules.sspro.p2.replay-only.m0.5.score\endcsname{90.6}
\expandafter\def\csname ppt@rev.pol.rules.sspro.p2.replay-only.m0.5.score\endcsname{$+$90.6}
\expandafter\def\csname val@rev.pol.rules.sspro.p2.replay-only.m1.keep\endcsname{1.000}
\expandafter\def\csname cnt@rev.pol.rules.sspro.p2.replay-only.m1.keep\endcsname{1}
\expandafter\def\csname pct@rev.pol.rules.sspro.p2.replay-only.m1.keep\endcsname{100.0}
\expandafter\def\csname ppt@rev.pol.rules.sspro.p2.replay-only.m1.keep\endcsname{$+$100.0}
\expandafter\def\csname val@rev.pol.rules.sspro.p2.replay-only.m1.req\endcsname{2.669}
\expandafter\def\csname cnt@rev.pol.rules.sspro.p2.replay-only.m1.req\endcsname{3}
\expandafter\def\csname pct@rev.pol.rules.sspro.p2.replay-only.m1.req\endcsname{266.9}
\expandafter\def\csname ppt@rev.pol.rules.sspro.p2.replay-only.m1.req\endcsname{$+$266.9}
\expandafter\def\csname val@rev.pol.rules.sspro.p2.replay-only.m1.score\endcsname{0.906}
\expandafter\def\csname cnt@rev.pol.rules.sspro.p2.replay-only.m1.score\endcsname{1}
\expandafter\def\csname pct@rev.pol.rules.sspro.p2.replay-only.m1.score\endcsname{90.6}
\expandafter\def\csname ppt@rev.pol.rules.sspro.p2.replay-only.m1.score\endcsname{$+$90.6}
\expandafter\def\csname val@rev.pol.rules.sspro.p2.replay-only.m2.keep\endcsname{1.000}
\expandafter\def\csname cnt@rev.pol.rules.sspro.p2.replay-only.m2.keep\endcsname{1}
\expandafter\def\csname pct@rev.pol.rules.sspro.p2.replay-only.m2.keep\endcsname{100.0}
\expandafter\def\csname ppt@rev.pol.rules.sspro.p2.replay-only.m2.keep\endcsname{$+$100.0}
\expandafter\def\csname val@rev.pol.rules.sspro.p2.replay-only.m2.req\endcsname{2.669}
\expandafter\def\csname cnt@rev.pol.rules.sspro.p2.replay-only.m2.req\endcsname{3}
\expandafter\def\csname pct@rev.pol.rules.sspro.p2.replay-only.m2.req\endcsname{266.9}
\expandafter\def\csname ppt@rev.pol.rules.sspro.p2.replay-only.m2.req\endcsname{$+$266.9}
\expandafter\def\csname val@rev.pol.rules.sspro.p2.replay-only.m2.score\endcsname{0.906}
\expandafter\def\csname cnt@rev.pol.rules.sspro.p2.replay-only.m2.score\endcsname{1}
\expandafter\def\csname pct@rev.pol.rules.sspro.p2.replay-only.m2.score\endcsname{90.6}
\expandafter\def\csname ppt@rev.pol.rules.sspro.p2.replay-only.m2.score\endcsname{$+$90.6}
\expandafter\def\csname val@rev.pol.rules.sspro.p2.replay-ppi-uniform@0.25.frozen.keep\endcsname{0.897}
\expandafter\def\csname cnt@rev.pol.rules.sspro.p2.replay-ppi-uniform@0.25.frozen.keep\endcsname{1}
\expandafter\def\csname pct@rev.pol.rules.sspro.p2.replay-ppi-uniform@0.25.frozen.keep\endcsname{89.7}
\expandafter\def\csname ppt@rev.pol.rules.sspro.p2.replay-ppi-uniform@0.25.frozen.keep\endcsname{$+$89.7}
\expandafter\def\csname val@rev.pol.rules.sspro.p2.replay-ppi-uniform@0.25.frozen.req\endcsname{2.446}
\expandafter\def\csname cnt@rev.pol.rules.sspro.p2.replay-ppi-uniform@0.25.frozen.req\endcsname{2}
\expandafter\def\csname pct@rev.pol.rules.sspro.p2.replay-ppi-uniform@0.25.frozen.req\endcsname{244.6}
\expandafter\def\csname ppt@rev.pol.rules.sspro.p2.replay-ppi-uniform@0.25.frozen.req\endcsname{$+$244.6}
\expandafter\def\csname val@rev.pol.rules.sspro.p2.replay-ppi-uniform@0.25.frozen.score\endcsname{0.891}
\expandafter\def\csname cnt@rev.pol.rules.sspro.p2.replay-ppi-uniform@0.25.frozen.score\endcsname{1}
\expandafter\def\csname pct@rev.pol.rules.sspro.p2.replay-ppi-uniform@0.25.frozen.score\endcsname{89.1}
\expandafter\def\csname ppt@rev.pol.rules.sspro.p2.replay-ppi-uniform@0.25.frozen.score\endcsname{$+$89.1}
\expandafter\def\csname val@rev.pol.rules.sspro.p2.replay-ppi-uniform@0.25.lam0.5.keep\endcsname{0.897}
\expandafter\def\csname cnt@rev.pol.rules.sspro.p2.replay-ppi-uniform@0.25.lam0.5.keep\endcsname{1}
\expandafter\def\csname pct@rev.pol.rules.sspro.p2.replay-ppi-uniform@0.25.lam0.5.keep\endcsname{89.7}
\expandafter\def\csname ppt@rev.pol.rules.sspro.p2.replay-ppi-uniform@0.25.lam0.5.keep\endcsname{$+$89.7}
\expandafter\def\csname val@rev.pol.rules.sspro.p2.replay-ppi-uniform@0.25.lam0.5.req\endcsname{2.446}
\expandafter\def\csname cnt@rev.pol.rules.sspro.p2.replay-ppi-uniform@0.25.lam0.5.req\endcsname{2}
\expandafter\def\csname pct@rev.pol.rules.sspro.p2.replay-ppi-uniform@0.25.lam0.5.req\endcsname{244.6}
\expandafter\def\csname ppt@rev.pol.rules.sspro.p2.replay-ppi-uniform@0.25.lam0.5.req\endcsname{$+$244.6}
\expandafter\def\csname val@rev.pol.rules.sspro.p2.replay-ppi-uniform@0.25.lam0.5.score\endcsname{0.891}
\expandafter\def\csname cnt@rev.pol.rules.sspro.p2.replay-ppi-uniform@0.25.lam0.5.score\endcsname{1}
\expandafter\def\csname pct@rev.pol.rules.sspro.p2.replay-ppi-uniform@0.25.lam0.5.score\endcsname{89.1}
\expandafter\def\csname ppt@rev.pol.rules.sspro.p2.replay-ppi-uniform@0.25.lam0.5.score\endcsname{$+$89.1}
\expandafter\def\csname val@rev.pol.rules.sspro.p2.replay-ppi-uniform@0.25.lam1.keep\endcsname{0.897}
\expandafter\def\csname cnt@rev.pol.rules.sspro.p2.replay-ppi-uniform@0.25.lam1.keep\endcsname{1}
\expandafter\def\csname pct@rev.pol.rules.sspro.p2.replay-ppi-uniform@0.25.lam1.keep\endcsname{89.7}
\expandafter\def\csname ppt@rev.pol.rules.sspro.p2.replay-ppi-uniform@0.25.lam1.keep\endcsname{$+$89.7}
\expandafter\def\csname val@rev.pol.rules.sspro.p2.replay-ppi-uniform@0.25.lam1.req\endcsname{2.446}
\expandafter\def\csname cnt@rev.pol.rules.sspro.p2.replay-ppi-uniform@0.25.lam1.req\endcsname{2}
\expandafter\def\csname pct@rev.pol.rules.sspro.p2.replay-ppi-uniform@0.25.lam1.req\endcsname{244.6}
\expandafter\def\csname ppt@rev.pol.rules.sspro.p2.replay-ppi-uniform@0.25.lam1.req\endcsname{$+$244.6}
\expandafter\def\csname val@rev.pol.rules.sspro.p2.replay-ppi-uniform@0.25.lam1.score\endcsname{0.891}
\expandafter\def\csname cnt@rev.pol.rules.sspro.p2.replay-ppi-uniform@0.25.lam1.score\endcsname{1}
\expandafter\def\csname pct@rev.pol.rules.sspro.p2.replay-ppi-uniform@0.25.lam1.score\endcsname{89.1}
\expandafter\def\csname ppt@rev.pol.rules.sspro.p2.replay-ppi-uniform@0.25.lam1.score\endcsname{$+$89.1}
\expandafter\def\csname val@rev.pol.rules.sspro.p2.replay-ppi-uniform@0.25.m0.5.keep\endcsname{0.897}
\expandafter\def\csname cnt@rev.pol.rules.sspro.p2.replay-ppi-uniform@0.25.m0.5.keep\endcsname{1}
\expandafter\def\csname pct@rev.pol.rules.sspro.p2.replay-ppi-uniform@0.25.m0.5.keep\endcsname{89.7}
\expandafter\def\csname ppt@rev.pol.rules.sspro.p2.replay-ppi-uniform@0.25.m0.5.keep\endcsname{$+$89.7}
\expandafter\def\csname val@rev.pol.rules.sspro.p2.replay-ppi-uniform@0.25.m0.5.req\endcsname{2.446}
\expandafter\def\csname cnt@rev.pol.rules.sspro.p2.replay-ppi-uniform@0.25.m0.5.req\endcsname{2}
\expandafter\def\csname pct@rev.pol.rules.sspro.p2.replay-ppi-uniform@0.25.m0.5.req\endcsname{244.6}
\expandafter\def\csname ppt@rev.pol.rules.sspro.p2.replay-ppi-uniform@0.25.m0.5.req\endcsname{$+$244.6}
\expandafter\def\csname val@rev.pol.rules.sspro.p2.replay-ppi-uniform@0.25.m0.5.score\endcsname{0.891}
\expandafter\def\csname cnt@rev.pol.rules.sspro.p2.replay-ppi-uniform@0.25.m0.5.score\endcsname{1}
\expandafter\def\csname pct@rev.pol.rules.sspro.p2.replay-ppi-uniform@0.25.m0.5.score\endcsname{89.1}
\expandafter\def\csname ppt@rev.pol.rules.sspro.p2.replay-ppi-uniform@0.25.m0.5.score\endcsname{$+$89.1}
\expandafter\def\csname val@rev.pol.rules.sspro.p2.replay-ppi-uniform@0.25.m1.keep\endcsname{0.897}
\expandafter\def\csname cnt@rev.pol.rules.sspro.p2.replay-ppi-uniform@0.25.m1.keep\endcsname{1}
\expandafter\def\csname pct@rev.pol.rules.sspro.p2.replay-ppi-uniform@0.25.m1.keep\endcsname{89.7}
\expandafter\def\csname ppt@rev.pol.rules.sspro.p2.replay-ppi-uniform@0.25.m1.keep\endcsname{$+$89.7}
\expandafter\def\csname val@rev.pol.rules.sspro.p2.replay-ppi-uniform@0.25.m1.req\endcsname{2.446}
\expandafter\def\csname cnt@rev.pol.rules.sspro.p2.replay-ppi-uniform@0.25.m1.req\endcsname{2}
\expandafter\def\csname pct@rev.pol.rules.sspro.p2.replay-ppi-uniform@0.25.m1.req\endcsname{244.6}
\expandafter\def\csname ppt@rev.pol.rules.sspro.p2.replay-ppi-uniform@0.25.m1.req\endcsname{$+$244.6}
\expandafter\def\csname val@rev.pol.rules.sspro.p2.replay-ppi-uniform@0.25.m1.score\endcsname{0.891}
\expandafter\def\csname cnt@rev.pol.rules.sspro.p2.replay-ppi-uniform@0.25.m1.score\endcsname{1}
\expandafter\def\csname pct@rev.pol.rules.sspro.p2.replay-ppi-uniform@0.25.m1.score\endcsname{89.1}
\expandafter\def\csname ppt@rev.pol.rules.sspro.p2.replay-ppi-uniform@0.25.m1.score\endcsname{$+$89.1}
\expandafter\def\csname val@rev.pol.rules.sspro.p2.replay-ppi-uniform@0.25.m2.keep\endcsname{0.897}
\expandafter\def\csname cnt@rev.pol.rules.sspro.p2.replay-ppi-uniform@0.25.m2.keep\endcsname{1}
\expandafter\def\csname pct@rev.pol.rules.sspro.p2.replay-ppi-uniform@0.25.m2.keep\endcsname{89.7}
\expandafter\def\csname ppt@rev.pol.rules.sspro.p2.replay-ppi-uniform@0.25.m2.keep\endcsname{$+$89.7}
\expandafter\def\csname val@rev.pol.rules.sspro.p2.replay-ppi-uniform@0.25.m2.req\endcsname{2.446}
\expandafter\def\csname cnt@rev.pol.rules.sspro.p2.replay-ppi-uniform@0.25.m2.req\endcsname{2}
\expandafter\def\csname pct@rev.pol.rules.sspro.p2.replay-ppi-uniform@0.25.m2.req\endcsname{244.6}
\expandafter\def\csname ppt@rev.pol.rules.sspro.p2.replay-ppi-uniform@0.25.m2.req\endcsname{$+$244.6}
\expandafter\def\csname val@rev.pol.rules.sspro.p2.replay-ppi-uniform@0.25.m2.score\endcsname{0.891}
\expandafter\def\csname cnt@rev.pol.rules.sspro.p2.replay-ppi-uniform@0.25.m2.score\endcsname{1}
\expandafter\def\csname pct@rev.pol.rules.sspro.p2.replay-ppi-uniform@0.25.m2.score\endcsname{89.1}
\expandafter\def\csname ppt@rev.pol.rules.sspro.p2.replay-ppi-uniform@0.25.m2.score\endcsname{$+$89.1}
\expandafter\def\csname val@rev.pol.rules.sspro.p2.suspend-all.keep\endcsname{0.000}
\expandafter\def\csname cnt@rev.pol.rules.sspro.p2.suspend-all.keep\endcsname{0}
\expandafter\def\csname pct@rev.pol.rules.sspro.p2.suspend-all.keep\endcsname{0.0}
\expandafter\def\csname ppt@rev.pol.rules.sspro.p2.suspend-all.keep\endcsname{0.0}
\expandafter\def\csname val@rev.pol.rules.sspro.p2.suspend-all.req\endcsname{1.000}
\expandafter\def\csname cnt@rev.pol.rules.sspro.p2.suspend-all.req\endcsname{1}
\expandafter\def\csname pct@rev.pol.rules.sspro.p2.suspend-all.req\endcsname{100.0}
\expandafter\def\csname ppt@rev.pol.rules.sspro.p2.suspend-all.req\endcsname{$+$100.0}
\expandafter\def\csname val@rev.pol.rules.sspro.p2.suspend-all.score\endcsname{0.760}
\expandafter\def\csname cnt@rev.pol.rules.sspro.p2.suspend-all.score\endcsname{1}
\expandafter\def\csname pct@rev.pol.rules.sspro.p2.suspend-all.score\endcsname{76.0}
\expandafter\def\csname ppt@rev.pol.rules.sspro.p2.suspend-all.score\endcsname{$+$76.0}
\expandafter\def\csname val@rev.pol.rules.sspro.p2.uniform-fresh@0.25.frozen.keep\endcsname{0.925}
\expandafter\def\csname cnt@rev.pol.rules.sspro.p2.uniform-fresh@0.25.frozen.keep\endcsname{1}
\expandafter\def\csname pct@rev.pol.rules.sspro.p2.uniform-fresh@0.25.frozen.keep\endcsname{92.5}
\expandafter\def\csname ppt@rev.pol.rules.sspro.p2.uniform-fresh@0.25.frozen.keep\endcsname{$+$92.5}
\expandafter\def\csname val@rev.pol.rules.sspro.p2.uniform-fresh@0.25.frozen.req\endcsname{2.524}
\expandafter\def\csname cnt@rev.pol.rules.sspro.p2.uniform-fresh@0.25.frozen.req\endcsname{3}
\expandafter\def\csname pct@rev.pol.rules.sspro.p2.uniform-fresh@0.25.frozen.req\endcsname{252.4}
\expandafter\def\csname ppt@rev.pol.rules.sspro.p2.uniform-fresh@0.25.frozen.req\endcsname{$+$252.4}
\expandafter\def\csname val@rev.pol.rules.sspro.p2.uniform-fresh@0.25.frozen.score\endcsname{0.895}
\expandafter\def\csname cnt@rev.pol.rules.sspro.p2.uniform-fresh@0.25.frozen.score\endcsname{1}
\expandafter\def\csname pct@rev.pol.rules.sspro.p2.uniform-fresh@0.25.frozen.score\endcsname{89.5}
\expandafter\def\csname ppt@rev.pol.rules.sspro.p2.uniform-fresh@0.25.frozen.score\endcsname{$+$89.5}
\expandafter\def\csname val@rev.pol.rules.sspro.p2.uniform-fresh@0.25.lam0.5.keep\endcsname{0.708}
\expandafter\def\csname cnt@rev.pol.rules.sspro.p2.uniform-fresh@0.25.lam0.5.keep\endcsname{1}
\expandafter\def\csname pct@rev.pol.rules.sspro.p2.uniform-fresh@0.25.lam0.5.keep\endcsname{70.8}
\expandafter\def\csname ppt@rev.pol.rules.sspro.p2.uniform-fresh@0.25.lam0.5.keep\endcsname{$+$70.8}
\expandafter\def\csname val@rev.pol.rules.sspro.p2.uniform-fresh@0.25.lam0.5.req\endcsname{2.160}
\expandafter\def\csname cnt@rev.pol.rules.sspro.p2.uniform-fresh@0.25.lam0.5.req\endcsname{2}
\expandafter\def\csname pct@rev.pol.rules.sspro.p2.uniform-fresh@0.25.lam0.5.req\endcsname{216.0}
\expandafter\def\csname ppt@rev.pol.rules.sspro.p2.uniform-fresh@0.25.lam0.5.req\endcsname{$+$216.0}
\expandafter\def\csname val@rev.pol.rules.sspro.p2.uniform-fresh@0.25.lam0.5.score\endcsname{0.863}
\expandafter\def\csname cnt@rev.pol.rules.sspro.p2.uniform-fresh@0.25.lam0.5.score\endcsname{1}
\expandafter\def\csname pct@rev.pol.rules.sspro.p2.uniform-fresh@0.25.lam0.5.score\endcsname{86.3}
\expandafter\def\csname ppt@rev.pol.rules.sspro.p2.uniform-fresh@0.25.lam0.5.score\endcsname{$+$86.3}
\expandafter\def\csname val@rev.pol.rules.sspro.p2.uniform-fresh@0.25.lam1.keep\endcsname{0.708}
\expandafter\def\csname cnt@rev.pol.rules.sspro.p2.uniform-fresh@0.25.lam1.keep\endcsname{1}
\expandafter\def\csname pct@rev.pol.rules.sspro.p2.uniform-fresh@0.25.lam1.keep\endcsname{70.8}
\expandafter\def\csname ppt@rev.pol.rules.sspro.p2.uniform-fresh@0.25.lam1.keep\endcsname{$+$70.8}
\expandafter\def\csname val@rev.pol.rules.sspro.p2.uniform-fresh@0.25.lam1.req\endcsname{2.160}
\expandafter\def\csname cnt@rev.pol.rules.sspro.p2.uniform-fresh@0.25.lam1.req\endcsname{2}
\expandafter\def\csname pct@rev.pol.rules.sspro.p2.uniform-fresh@0.25.lam1.req\endcsname{216.0}
\expandafter\def\csname ppt@rev.pol.rules.sspro.p2.uniform-fresh@0.25.lam1.req\endcsname{$+$216.0}
\expandafter\def\csname val@rev.pol.rules.sspro.p2.uniform-fresh@0.25.lam1.score\endcsname{0.863}
\expandafter\def\csname cnt@rev.pol.rules.sspro.p2.uniform-fresh@0.25.lam1.score\endcsname{1}
\expandafter\def\csname pct@rev.pol.rules.sspro.p2.uniform-fresh@0.25.lam1.score\endcsname{86.3}
\expandafter\def\csname ppt@rev.pol.rules.sspro.p2.uniform-fresh@0.25.lam1.score\endcsname{$+$86.3}
\expandafter\def\csname val@rev.pol.rules.sspro.p2.uniform-fresh@0.25.m0.5.keep\endcsname{0.708}
\expandafter\def\csname cnt@rev.pol.rules.sspro.p2.uniform-fresh@0.25.m0.5.keep\endcsname{1}
\expandafter\def\csname pct@rev.pol.rules.sspro.p2.uniform-fresh@0.25.m0.5.keep\endcsname{70.8}
\expandafter\def\csname ppt@rev.pol.rules.sspro.p2.uniform-fresh@0.25.m0.5.keep\endcsname{$+$70.8}
\expandafter\def\csname val@rev.pol.rules.sspro.p2.uniform-fresh@0.25.m0.5.req\endcsname{2.160}
\expandafter\def\csname cnt@rev.pol.rules.sspro.p2.uniform-fresh@0.25.m0.5.req\endcsname{2}
\expandafter\def\csname pct@rev.pol.rules.sspro.p2.uniform-fresh@0.25.m0.5.req\endcsname{216.0}
\expandafter\def\csname ppt@rev.pol.rules.sspro.p2.uniform-fresh@0.25.m0.5.req\endcsname{$+$216.0}
\expandafter\def\csname val@rev.pol.rules.sspro.p2.uniform-fresh@0.25.m0.5.score\endcsname{0.863}
\expandafter\def\csname cnt@rev.pol.rules.sspro.p2.uniform-fresh@0.25.m0.5.score\endcsname{1}
\expandafter\def\csname pct@rev.pol.rules.sspro.p2.uniform-fresh@0.25.m0.5.score\endcsname{86.3}
\expandafter\def\csname ppt@rev.pol.rules.sspro.p2.uniform-fresh@0.25.m0.5.score\endcsname{$+$86.3}
\expandafter\def\csname val@rev.pol.rules.sspro.p2.uniform-fresh@0.25.m1.keep\endcsname{0.708}
\expandafter\def\csname cnt@rev.pol.rules.sspro.p2.uniform-fresh@0.25.m1.keep\endcsname{1}
\expandafter\def\csname pct@rev.pol.rules.sspro.p2.uniform-fresh@0.25.m1.keep\endcsname{70.8}
\expandafter\def\csname ppt@rev.pol.rules.sspro.p2.uniform-fresh@0.25.m1.keep\endcsname{$+$70.8}
\expandafter\def\csname val@rev.pol.rules.sspro.p2.uniform-fresh@0.25.m1.req\endcsname{2.160}
\expandafter\def\csname cnt@rev.pol.rules.sspro.p2.uniform-fresh@0.25.m1.req\endcsname{2}
\expandafter\def\csname pct@rev.pol.rules.sspro.p2.uniform-fresh@0.25.m1.req\endcsname{216.0}
\expandafter\def\csname ppt@rev.pol.rules.sspro.p2.uniform-fresh@0.25.m1.req\endcsname{$+$216.0}
\expandafter\def\csname val@rev.pol.rules.sspro.p2.uniform-fresh@0.25.m1.score\endcsname{0.863}
\expandafter\def\csname cnt@rev.pol.rules.sspro.p2.uniform-fresh@0.25.m1.score\endcsname{1}
\expandafter\def\csname pct@rev.pol.rules.sspro.p2.uniform-fresh@0.25.m1.score\endcsname{86.3}
\expandafter\def\csname ppt@rev.pol.rules.sspro.p2.uniform-fresh@0.25.m1.score\endcsname{$+$86.3}
\expandafter\def\csname val@rev.pol.rules.sspro.p2.uniform-fresh@0.25.m2.keep\endcsname{0.708}
\expandafter\def\csname cnt@rev.pol.rules.sspro.p2.uniform-fresh@0.25.m2.keep\endcsname{1}
\expandafter\def\csname pct@rev.pol.rules.sspro.p2.uniform-fresh@0.25.m2.keep\endcsname{70.8}
\expandafter\def\csname ppt@rev.pol.rules.sspro.p2.uniform-fresh@0.25.m2.keep\endcsname{$+$70.8}
\expandafter\def\csname val@rev.pol.rules.sspro.p2.uniform-fresh@0.25.m2.req\endcsname{2.160}
\expandafter\def\csname cnt@rev.pol.rules.sspro.p2.uniform-fresh@0.25.m2.req\endcsname{2}
\expandafter\def\csname pct@rev.pol.rules.sspro.p2.uniform-fresh@0.25.m2.req\endcsname{216.0}
\expandafter\def\csname ppt@rev.pol.rules.sspro.p2.uniform-fresh@0.25.m2.req\endcsname{$+$216.0}
\expandafter\def\csname val@rev.pol.rules.sspro.p2.uniform-fresh@0.25.m2.score\endcsname{0.863}
\expandafter\def\csname cnt@rev.pol.rules.sspro.p2.uniform-fresh@0.25.m2.score\endcsname{1}
\expandafter\def\csname pct@rev.pol.rules.sspro.p2.uniform-fresh@0.25.m2.score\endcsname{86.3}
\expandafter\def\csname ppt@rev.pol.rules.sspro.p2.uniform-fresh@0.25.m2.score\endcsname{$+$86.3}
\expandafter\def\csname val@rev.pol.rules.uiv.p1.full-recert.frozen.keep\endcsname{0.500}
\expandafter\def\csname cnt@rev.pol.rules.uiv.p1.full-recert.frozen.keep\endcsname{0}
\expandafter\def\csname pct@rev.pol.rules.uiv.p1.full-recert.frozen.keep\endcsname{50.0}
\expandafter\def\csname ppt@rev.pol.rules.uiv.p1.full-recert.frozen.keep\endcsname{$+$50.0}
\expandafter\def\csname val@rev.pol.rules.uiv.p1.full-recert.frozen.req\endcsname{1.757}
\expandafter\def\csname cnt@rev.pol.rules.uiv.p1.full-recert.frozen.req\endcsname{2}
\expandafter\def\csname pct@rev.pol.rules.uiv.p1.full-recert.frozen.req\endcsname{175.7}
\expandafter\def\csname ppt@rev.pol.rules.uiv.p1.full-recert.frozen.req\endcsname{$+$175.7}
\expandafter\def\csname val@rev.pol.rules.uiv.p1.full-recert.frozen.score\endcsname{0.793}
\expandafter\def\csname cnt@rev.pol.rules.uiv.p1.full-recert.frozen.score\endcsname{1}
\expandafter\def\csname pct@rev.pol.rules.uiv.p1.full-recert.frozen.score\endcsname{79.3}
\expandafter\def\csname ppt@rev.pol.rules.uiv.p1.full-recert.frozen.score\endcsname{$+$79.3}
\expandafter\def\csname val@rev.pol.rules.uiv.p1.full-recert.lam0.5.keep\endcsname{0.250}
\expandafter\def\csname cnt@rev.pol.rules.uiv.p1.full-recert.lam0.5.keep\endcsname{0}
\expandafter\def\csname pct@rev.pol.rules.uiv.p1.full-recert.lam0.5.keep\endcsname{25.0}
\expandafter\def\csname ppt@rev.pol.rules.uiv.p1.full-recert.lam0.5.keep\endcsname{$+$25.0}
\expandafter\def\csname val@rev.pol.rules.uiv.p1.full-recert.lam0.5.req\endcsname{1.252}
\expandafter\def\csname cnt@rev.pol.rules.uiv.p1.full-recert.lam0.5.req\endcsname{1}
\expandafter\def\csname pct@rev.pol.rules.uiv.p1.full-recert.lam0.5.req\endcsname{125.2}
\expandafter\def\csname ppt@rev.pol.rules.uiv.p1.full-recert.lam0.5.req\endcsname{$+$125.2}
\expandafter\def\csname val@rev.pol.rules.uiv.p1.full-recert.lam0.5.score\endcsname{0.819}
\expandafter\def\csname cnt@rev.pol.rules.uiv.p1.full-recert.lam0.5.score\endcsname{1}
\expandafter\def\csname pct@rev.pol.rules.uiv.p1.full-recert.lam0.5.score\endcsname{81.9}
\expandafter\def\csname ppt@rev.pol.rules.uiv.p1.full-recert.lam0.5.score\endcsname{$+$81.9}
\expandafter\def\csname val@rev.pol.rules.uiv.p1.full-recert.lam1.keep\endcsname{0.250}
\expandafter\def\csname cnt@rev.pol.rules.uiv.p1.full-recert.lam1.keep\endcsname{0}
\expandafter\def\csname pct@rev.pol.rules.uiv.p1.full-recert.lam1.keep\endcsname{25.0}
\expandafter\def\csname ppt@rev.pol.rules.uiv.p1.full-recert.lam1.keep\endcsname{$+$25.0}
\expandafter\def\csname val@rev.pol.rules.uiv.p1.full-recert.lam1.req\endcsname{1.252}
\expandafter\def\csname cnt@rev.pol.rules.uiv.p1.full-recert.lam1.req\endcsname{1}
\expandafter\def\csname pct@rev.pol.rules.uiv.p1.full-recert.lam1.req\endcsname{125.2}
\expandafter\def\csname ppt@rev.pol.rules.uiv.p1.full-recert.lam1.req\endcsname{$+$125.2}
\expandafter\def\csname val@rev.pol.rules.uiv.p1.full-recert.lam1.score\endcsname{0.819}
\expandafter\def\csname cnt@rev.pol.rules.uiv.p1.full-recert.lam1.score\endcsname{1}
\expandafter\def\csname pct@rev.pol.rules.uiv.p1.full-recert.lam1.score\endcsname{81.9}
\expandafter\def\csname ppt@rev.pol.rules.uiv.p1.full-recert.lam1.score\endcsname{$+$81.9}
\expandafter\def\csname val@rev.pol.rules.uiv.p1.full-recert.m0.5.keep\endcsname{0.250}
\expandafter\def\csname cnt@rev.pol.rules.uiv.p1.full-recert.m0.5.keep\endcsname{0}
\expandafter\def\csname pct@rev.pol.rules.uiv.p1.full-recert.m0.5.keep\endcsname{25.0}
\expandafter\def\csname ppt@rev.pol.rules.uiv.p1.full-recert.m0.5.keep\endcsname{$+$25.0}
\expandafter\def\csname val@rev.pol.rules.uiv.p1.full-recert.m0.5.req\endcsname{1.252}
\expandafter\def\csname cnt@rev.pol.rules.uiv.p1.full-recert.m0.5.req\endcsname{1}
\expandafter\def\csname pct@rev.pol.rules.uiv.p1.full-recert.m0.5.req\endcsname{125.2}
\expandafter\def\csname ppt@rev.pol.rules.uiv.p1.full-recert.m0.5.req\endcsname{$+$125.2}
\expandafter\def\csname val@rev.pol.rules.uiv.p1.full-recert.m0.5.score\endcsname{0.819}
\expandafter\def\csname cnt@rev.pol.rules.uiv.p1.full-recert.m0.5.score\endcsname{1}
\expandafter\def\csname pct@rev.pol.rules.uiv.p1.full-recert.m0.5.score\endcsname{81.9}
\expandafter\def\csname ppt@rev.pol.rules.uiv.p1.full-recert.m0.5.score\endcsname{$+$81.9}
\expandafter\def\csname val@rev.pol.rules.uiv.p1.full-recert.m1.keep\endcsname{0.250}
\expandafter\def\csname cnt@rev.pol.rules.uiv.p1.full-recert.m1.keep\endcsname{0}
\expandafter\def\csname pct@rev.pol.rules.uiv.p1.full-recert.m1.keep\endcsname{25.0}
\expandafter\def\csname ppt@rev.pol.rules.uiv.p1.full-recert.m1.keep\endcsname{$+$25.0}
\expandafter\def\csname val@rev.pol.rules.uiv.p1.full-recert.m1.req\endcsname{1.252}
\expandafter\def\csname cnt@rev.pol.rules.uiv.p1.full-recert.m1.req\endcsname{1}
\expandafter\def\csname pct@rev.pol.rules.uiv.p1.full-recert.m1.req\endcsname{125.2}
\expandafter\def\csname ppt@rev.pol.rules.uiv.p1.full-recert.m1.req\endcsname{$+$125.2}
\expandafter\def\csname val@rev.pol.rules.uiv.p1.full-recert.m1.score\endcsname{0.819}
\expandafter\def\csname cnt@rev.pol.rules.uiv.p1.full-recert.m1.score\endcsname{1}
\expandafter\def\csname pct@rev.pol.rules.uiv.p1.full-recert.m1.score\endcsname{81.9}
\expandafter\def\csname ppt@rev.pol.rules.uiv.p1.full-recert.m1.score\endcsname{$+$81.9}
\expandafter\def\csname val@rev.pol.rules.uiv.p1.full-recert.m2.keep\endcsname{0.000}
\expandafter\def\csname cnt@rev.pol.rules.uiv.p1.full-recert.m2.keep\endcsname{0}
\expandafter\def\csname pct@rev.pol.rules.uiv.p1.full-recert.m2.keep\endcsname{0.0}
\expandafter\def\csname ppt@rev.pol.rules.uiv.p1.full-recert.m2.keep\endcsname{0.0}
\expandafter\def\csname val@rev.pol.rules.uiv.p1.full-recert.m2.req\endcsname{1.000}
\expandafter\def\csname cnt@rev.pol.rules.uiv.p1.full-recert.m2.req\endcsname{1}
\expandafter\def\csname pct@rev.pol.rules.uiv.p1.full-recert.m2.req\endcsname{100.0}
\expandafter\def\csname ppt@rev.pol.rules.uiv.p1.full-recert.m2.req\endcsname{$+$100.0}
\expandafter\def\csname val@rev.pol.rules.uiv.p1.full-recert.m2.score\endcsname{0.833}
\expandafter\def\csname cnt@rev.pol.rules.uiv.p1.full-recert.m2.score\endcsname{1}
\expandafter\def\csname pct@rev.pol.rules.uiv.p1.full-recert.m2.score\endcsname{83.3}
\expandafter\def\csname ppt@rev.pol.rules.uiv.p1.full-recert.m2.score\endcsname{$+$83.3}
\expandafter\def\csname val@rev.pol.rules.uiv.p1.generic-ppi-source@0.25.frozen.keep\endcsname{0.496}
\expandafter\def\csname cnt@rev.pol.rules.uiv.p1.generic-ppi-source@0.25.frozen.keep\endcsname{0}
\expandafter\def\csname pct@rev.pol.rules.uiv.p1.generic-ppi-source@0.25.frozen.keep\endcsname{49.6}
\expandafter\def\csname ppt@rev.pol.rules.uiv.p1.generic-ppi-source@0.25.frozen.keep\endcsname{$+$49.6}
\expandafter\def\csname val@rev.pol.rules.uiv.p1.generic-ppi-source@0.25.frozen.req\endcsname{1.795}
\expandafter\def\csname cnt@rev.pol.rules.uiv.p1.generic-ppi-source@0.25.frozen.req\endcsname{2}
\expandafter\def\csname pct@rev.pol.rules.uiv.p1.generic-ppi-source@0.25.frozen.req\endcsname{179.5}
\expandafter\def\csname ppt@rev.pol.rules.uiv.p1.generic-ppi-source@0.25.frozen.req\endcsname{$+$179.5}
\expandafter\def\csname val@rev.pol.rules.uiv.p1.generic-ppi-source@0.25.frozen.score\endcsname{0.797}
\expandafter\def\csname cnt@rev.pol.rules.uiv.p1.generic-ppi-source@0.25.frozen.score\endcsname{1}
\expandafter\def\csname pct@rev.pol.rules.uiv.p1.generic-ppi-source@0.25.frozen.score\endcsname{79.7}
\expandafter\def\csname ppt@rev.pol.rules.uiv.p1.generic-ppi-source@0.25.frozen.score\endcsname{$+$79.7}
\expandafter\def\csname val@rev.pol.rules.uiv.p1.generic-ppi-source@0.25.lam0.5.keep\endcsname{0.444}
\expandafter\def\csname cnt@rev.pol.rules.uiv.p1.generic-ppi-source@0.25.lam0.5.keep\endcsname{0}
\expandafter\def\csname pct@rev.pol.rules.uiv.p1.generic-ppi-source@0.25.lam0.5.keep\endcsname{44.4}
\expandafter\def\csname ppt@rev.pol.rules.uiv.p1.generic-ppi-source@0.25.lam0.5.keep\endcsname{$+$44.4}
\expandafter\def\csname val@rev.pol.rules.uiv.p1.generic-ppi-source@0.25.lam0.5.req\endcsname{1.691}
\expandafter\def\csname cnt@rev.pol.rules.uiv.p1.generic-ppi-source@0.25.lam0.5.req\endcsname{2}
\expandafter\def\csname pct@rev.pol.rules.uiv.p1.generic-ppi-source@0.25.lam0.5.req\endcsname{169.1}
\expandafter\def\csname ppt@rev.pol.rules.uiv.p1.generic-ppi-source@0.25.lam0.5.req\endcsname{$+$169.1}
\expandafter\def\csname val@rev.pol.rules.uiv.p1.generic-ppi-source@0.25.lam0.5.score\endcsname{0.802}
\expandafter\def\csname cnt@rev.pol.rules.uiv.p1.generic-ppi-source@0.25.lam0.5.score\endcsname{1}
\expandafter\def\csname pct@rev.pol.rules.uiv.p1.generic-ppi-source@0.25.lam0.5.score\endcsname{80.2}
\expandafter\def\csname ppt@rev.pol.rules.uiv.p1.generic-ppi-source@0.25.lam0.5.score\endcsname{$+$80.2}
\expandafter\def\csname val@rev.pol.rules.uiv.p1.generic-ppi-source@0.25.lam1.keep\endcsname{0.404}
\expandafter\def\csname cnt@rev.pol.rules.uiv.p1.generic-ppi-source@0.25.lam1.keep\endcsname{0}
\expandafter\def\csname pct@rev.pol.rules.uiv.p1.generic-ppi-source@0.25.lam1.keep\endcsname{40.4}
\expandafter\def\csname ppt@rev.pol.rules.uiv.p1.generic-ppi-source@0.25.lam1.keep\endcsname{$+$40.4}
\expandafter\def\csname val@rev.pol.rules.uiv.p1.generic-ppi-source@0.25.lam1.req\endcsname{1.585}
\expandafter\def\csname cnt@rev.pol.rules.uiv.p1.generic-ppi-source@0.25.lam1.req\endcsname{2}
\expandafter\def\csname pct@rev.pol.rules.uiv.p1.generic-ppi-source@0.25.lam1.req\endcsname{158.5}
\expandafter\def\csname ppt@rev.pol.rules.uiv.p1.generic-ppi-source@0.25.lam1.req\endcsname{$+$158.5}
\expandafter\def\csname val@rev.pol.rules.uiv.p1.generic-ppi-source@0.25.lam1.score\endcsname{0.805}
\expandafter\def\csname cnt@rev.pol.rules.uiv.p1.generic-ppi-source@0.25.lam1.score\endcsname{1}
\expandafter\def\csname pct@rev.pol.rules.uiv.p1.generic-ppi-source@0.25.lam1.score\endcsname{80.5}
\expandafter\def\csname ppt@rev.pol.rules.uiv.p1.generic-ppi-source@0.25.lam1.score\endcsname{$+$80.5}
\expandafter\def\csname val@rev.pol.rules.uiv.p1.generic-ppi-source@0.25.m0.5.keep\endcsname{0.444}
\expandafter\def\csname cnt@rev.pol.rules.uiv.p1.generic-ppi-source@0.25.m0.5.keep\endcsname{0}
\expandafter\def\csname pct@rev.pol.rules.uiv.p1.generic-ppi-source@0.25.m0.5.keep\endcsname{44.4}
\expandafter\def\csname ppt@rev.pol.rules.uiv.p1.generic-ppi-source@0.25.m0.5.keep\endcsname{$+$44.4}
\expandafter\def\csname val@rev.pol.rules.uiv.p1.generic-ppi-source@0.25.m0.5.req\endcsname{1.691}
\expandafter\def\csname cnt@rev.pol.rules.uiv.p1.generic-ppi-source@0.25.m0.5.req\endcsname{2}
\expandafter\def\csname pct@rev.pol.rules.uiv.p1.generic-ppi-source@0.25.m0.5.req\endcsname{169.1}
\expandafter\def\csname ppt@rev.pol.rules.uiv.p1.generic-ppi-source@0.25.m0.5.req\endcsname{$+$169.1}
\expandafter\def\csname val@rev.pol.rules.uiv.p1.generic-ppi-source@0.25.m0.5.score\endcsname{0.802}
\expandafter\def\csname cnt@rev.pol.rules.uiv.p1.generic-ppi-source@0.25.m0.5.score\endcsname{1}
\expandafter\def\csname pct@rev.pol.rules.uiv.p1.generic-ppi-source@0.25.m0.5.score\endcsname{80.2}
\expandafter\def\csname ppt@rev.pol.rules.uiv.p1.generic-ppi-source@0.25.m0.5.score\endcsname{$+$80.2}
\expandafter\def\csname val@rev.pol.rules.uiv.p1.generic-ppi-source@0.25.m1.keep\endcsname{0.444}
\expandafter\def\csname cnt@rev.pol.rules.uiv.p1.generic-ppi-source@0.25.m1.keep\endcsname{0}
\expandafter\def\csname pct@rev.pol.rules.uiv.p1.generic-ppi-source@0.25.m1.keep\endcsname{44.4}
\expandafter\def\csname ppt@rev.pol.rules.uiv.p1.generic-ppi-source@0.25.m1.keep\endcsname{$+$44.4}
\expandafter\def\csname val@rev.pol.rules.uiv.p1.generic-ppi-source@0.25.m1.req\endcsname{1.691}
\expandafter\def\csname cnt@rev.pol.rules.uiv.p1.generic-ppi-source@0.25.m1.req\endcsname{2}
\expandafter\def\csname pct@rev.pol.rules.uiv.p1.generic-ppi-source@0.25.m1.req\endcsname{169.1}
\expandafter\def\csname ppt@rev.pol.rules.uiv.p1.generic-ppi-source@0.25.m1.req\endcsname{$+$169.1}
\expandafter\def\csname val@rev.pol.rules.uiv.p1.generic-ppi-source@0.25.m1.score\endcsname{0.802}
\expandafter\def\csname cnt@rev.pol.rules.uiv.p1.generic-ppi-source@0.25.m1.score\endcsname{1}
\expandafter\def\csname pct@rev.pol.rules.uiv.p1.generic-ppi-source@0.25.m1.score\endcsname{80.2}
\expandafter\def\csname ppt@rev.pol.rules.uiv.p1.generic-ppi-source@0.25.m1.score\endcsname{$+$80.2}
\expandafter\def\csname val@rev.pol.rules.uiv.p1.generic-ppi-source@0.25.m2.keep\endcsname{0.341}
\expandafter\def\csname cnt@rev.pol.rules.uiv.p1.generic-ppi-source@0.25.m2.keep\endcsname{0}
\expandafter\def\csname pct@rev.pol.rules.uiv.p1.generic-ppi-source@0.25.m2.keep\endcsname{34.1}
\expandafter\def\csname ppt@rev.pol.rules.uiv.p1.generic-ppi-source@0.25.m2.keep\endcsname{$+$34.1}
\expandafter\def\csname val@rev.pol.rules.uiv.p1.generic-ppi-source@0.25.m2.req\endcsname{1.587}
\expandafter\def\csname cnt@rev.pol.rules.uiv.p1.generic-ppi-source@0.25.m2.req\endcsname{2}
\expandafter\def\csname pct@rev.pol.rules.uiv.p1.generic-ppi-source@0.25.m2.req\endcsname{158.7}
\expandafter\def\csname ppt@rev.pol.rules.uiv.p1.generic-ppi-source@0.25.m2.req\endcsname{$+$158.7}
\expandafter\def\csname val@rev.pol.rules.uiv.p1.generic-ppi-source@0.25.m2.score\endcsname{0.808}
\expandafter\def\csname cnt@rev.pol.rules.uiv.p1.generic-ppi-source@0.25.m2.score\endcsname{1}
\expandafter\def\csname pct@rev.pol.rules.uiv.p1.generic-ppi-source@0.25.m2.score\endcsname{80.8}
\expandafter\def\csname ppt@rev.pol.rules.uiv.p1.generic-ppi-source@0.25.m2.score\endcsname{$+$80.8}
\expandafter\def\csname val@rev.pol.rules.uiv.p1.keep-all.keep\endcsname{1.000}
\expandafter\def\csname cnt@rev.pol.rules.uiv.p1.keep-all.keep\endcsname{1}
\expandafter\def\csname pct@rev.pol.rules.uiv.p1.keep-all.keep\endcsname{100.0}
\expandafter\def\csname ppt@rev.pol.rules.uiv.p1.keep-all.keep\endcsname{$+$100.0}
\expandafter\def\csname val@rev.pol.rules.uiv.p1.keep-all.req\endcsname{2.672}
\expandafter\def\csname cnt@rev.pol.rules.uiv.p1.keep-all.req\endcsname{3}
\expandafter\def\csname pct@rev.pol.rules.uiv.p1.keep-all.req\endcsname{267.2}
\expandafter\def\csname ppt@rev.pol.rules.uiv.p1.keep-all.req\endcsname{$+$267.2}
\expandafter\def\csname val@rev.pol.rules.uiv.p1.keep-all.score\endcsname{0.760}
\expandafter\def\csname cnt@rev.pol.rules.uiv.p1.keep-all.score\endcsname{1}
\expandafter\def\csname pct@rev.pol.rules.uiv.p1.keep-all.score\endcsname{76.0}
\expandafter\def\csname ppt@rev.pol.rules.uiv.p1.keep-all.score\endcsname{$+$76.0}
\expandafter\def\csname val@rev.pol.rules.uiv.p1.old-decision.frozen.keep\endcsname{0.000}
\expandafter\def\csname cnt@rev.pol.rules.uiv.p1.old-decision.frozen.keep\endcsname{0}
\expandafter\def\csname pct@rev.pol.rules.uiv.p1.old-decision.frozen.keep\endcsname{0.0}
\expandafter\def\csname ppt@rev.pol.rules.uiv.p1.old-decision.frozen.keep\endcsname{0.0}
\expandafter\def\csname val@rev.pol.rules.uiv.p1.old-decision.frozen.req\endcsname{1.000}
\expandafter\def\csname cnt@rev.pol.rules.uiv.p1.old-decision.frozen.req\endcsname{1}
\expandafter\def\csname pct@rev.pol.rules.uiv.p1.old-decision.frozen.req\endcsname{100.0}
\expandafter\def\csname ppt@rev.pol.rules.uiv.p1.old-decision.frozen.req\endcsname{$+$100.0}
\expandafter\def\csname val@rev.pol.rules.uiv.p1.old-decision.frozen.score\endcsname{0.833}
\expandafter\def\csname cnt@rev.pol.rules.uiv.p1.old-decision.frozen.score\endcsname{1}
\expandafter\def\csname pct@rev.pol.rules.uiv.p1.old-decision.frozen.score\endcsname{83.3}
\expandafter\def\csname ppt@rev.pol.rules.uiv.p1.old-decision.frozen.score\endcsname{$+$83.3}
\expandafter\def\csname val@rev.pol.rules.uiv.p1.old-decision.lam0.5.keep\endcsname{0.000}
\expandafter\def\csname cnt@rev.pol.rules.uiv.p1.old-decision.lam0.5.keep\endcsname{0}
\expandafter\def\csname pct@rev.pol.rules.uiv.p1.old-decision.lam0.5.keep\endcsname{0.0}
\expandafter\def\csname ppt@rev.pol.rules.uiv.p1.old-decision.lam0.5.keep\endcsname{0.0}
\expandafter\def\csname val@rev.pol.rules.uiv.p1.old-decision.lam0.5.req\endcsname{1.000}
\expandafter\def\csname cnt@rev.pol.rules.uiv.p1.old-decision.lam0.5.req\endcsname{1}
\expandafter\def\csname pct@rev.pol.rules.uiv.p1.old-decision.lam0.5.req\endcsname{100.0}
\expandafter\def\csname ppt@rev.pol.rules.uiv.p1.old-decision.lam0.5.req\endcsname{$+$100.0}
\expandafter\def\csname val@rev.pol.rules.uiv.p1.old-decision.lam0.5.score\endcsname{0.833}
\expandafter\def\csname cnt@rev.pol.rules.uiv.p1.old-decision.lam0.5.score\endcsname{1}
\expandafter\def\csname pct@rev.pol.rules.uiv.p1.old-decision.lam0.5.score\endcsname{83.3}
\expandafter\def\csname ppt@rev.pol.rules.uiv.p1.old-decision.lam0.5.score\endcsname{$+$83.3}
\expandafter\def\csname val@rev.pol.rules.uiv.p1.old-decision.lam1.keep\endcsname{0.000}
\expandafter\def\csname cnt@rev.pol.rules.uiv.p1.old-decision.lam1.keep\endcsname{0}
\expandafter\def\csname pct@rev.pol.rules.uiv.p1.old-decision.lam1.keep\endcsname{0.0}
\expandafter\def\csname ppt@rev.pol.rules.uiv.p1.old-decision.lam1.keep\endcsname{0.0}
\expandafter\def\csname val@rev.pol.rules.uiv.p1.old-decision.lam1.req\endcsname{1.000}
\expandafter\def\csname cnt@rev.pol.rules.uiv.p1.old-decision.lam1.req\endcsname{1}
\expandafter\def\csname pct@rev.pol.rules.uiv.p1.old-decision.lam1.req\endcsname{100.0}
\expandafter\def\csname ppt@rev.pol.rules.uiv.p1.old-decision.lam1.req\endcsname{$+$100.0}
\expandafter\def\csname val@rev.pol.rules.uiv.p1.old-decision.lam1.score\endcsname{0.833}
\expandafter\def\csname cnt@rev.pol.rules.uiv.p1.old-decision.lam1.score\endcsname{1}
\expandafter\def\csname pct@rev.pol.rules.uiv.p1.old-decision.lam1.score\endcsname{83.3}
\expandafter\def\csname ppt@rev.pol.rules.uiv.p1.old-decision.lam1.score\endcsname{$+$83.3}
\expandafter\def\csname val@rev.pol.rules.uiv.p1.old-decision.m0.5.keep\endcsname{0.000}
\expandafter\def\csname cnt@rev.pol.rules.uiv.p1.old-decision.m0.5.keep\endcsname{0}
\expandafter\def\csname pct@rev.pol.rules.uiv.p1.old-decision.m0.5.keep\endcsname{0.0}
\expandafter\def\csname ppt@rev.pol.rules.uiv.p1.old-decision.m0.5.keep\endcsname{0.0}
\expandafter\def\csname val@rev.pol.rules.uiv.p1.old-decision.m0.5.req\endcsname{1.000}
\expandafter\def\csname cnt@rev.pol.rules.uiv.p1.old-decision.m0.5.req\endcsname{1}
\expandafter\def\csname pct@rev.pol.rules.uiv.p1.old-decision.m0.5.req\endcsname{100.0}
\expandafter\def\csname ppt@rev.pol.rules.uiv.p1.old-decision.m0.5.req\endcsname{$+$100.0}
\expandafter\def\csname val@rev.pol.rules.uiv.p1.old-decision.m0.5.score\endcsname{0.833}
\expandafter\def\csname cnt@rev.pol.rules.uiv.p1.old-decision.m0.5.score\endcsname{1}
\expandafter\def\csname pct@rev.pol.rules.uiv.p1.old-decision.m0.5.score\endcsname{83.3}
\expandafter\def\csname ppt@rev.pol.rules.uiv.p1.old-decision.m0.5.score\endcsname{$+$83.3}
\expandafter\def\csname val@rev.pol.rules.uiv.p1.old-decision.m1.keep\endcsname{0.000}
\expandafter\def\csname cnt@rev.pol.rules.uiv.p1.old-decision.m1.keep\endcsname{0}
\expandafter\def\csname pct@rev.pol.rules.uiv.p1.old-decision.m1.keep\endcsname{0.0}
\expandafter\def\csname ppt@rev.pol.rules.uiv.p1.old-decision.m1.keep\endcsname{0.0}
\expandafter\def\csname val@rev.pol.rules.uiv.p1.old-decision.m1.req\endcsname{1.000}
\expandafter\def\csname cnt@rev.pol.rules.uiv.p1.old-decision.m1.req\endcsname{1}
\expandafter\def\csname pct@rev.pol.rules.uiv.p1.old-decision.m1.req\endcsname{100.0}
\expandafter\def\csname ppt@rev.pol.rules.uiv.p1.old-decision.m1.req\endcsname{$+$100.0}
\expandafter\def\csname val@rev.pol.rules.uiv.p1.old-decision.m1.score\endcsname{0.833}
\expandafter\def\csname cnt@rev.pol.rules.uiv.p1.old-decision.m1.score\endcsname{1}
\expandafter\def\csname pct@rev.pol.rules.uiv.p1.old-decision.m1.score\endcsname{83.3}
\expandafter\def\csname ppt@rev.pol.rules.uiv.p1.old-decision.m1.score\endcsname{$+$83.3}
\expandafter\def\csname val@rev.pol.rules.uiv.p1.old-decision.m2.keep\endcsname{0.000}
\expandafter\def\csname cnt@rev.pol.rules.uiv.p1.old-decision.m2.keep\endcsname{0}
\expandafter\def\csname pct@rev.pol.rules.uiv.p1.old-decision.m2.keep\endcsname{0.0}
\expandafter\def\csname ppt@rev.pol.rules.uiv.p1.old-decision.m2.keep\endcsname{0.0}
\expandafter\def\csname val@rev.pol.rules.uiv.p1.old-decision.m2.req\endcsname{1.000}
\expandafter\def\csname cnt@rev.pol.rules.uiv.p1.old-decision.m2.req\endcsname{1}
\expandafter\def\csname pct@rev.pol.rules.uiv.p1.old-decision.m2.req\endcsname{100.0}
\expandafter\def\csname ppt@rev.pol.rules.uiv.p1.old-decision.m2.req\endcsname{$+$100.0}
\expandafter\def\csname val@rev.pol.rules.uiv.p1.old-decision.m2.score\endcsname{0.833}
\expandafter\def\csname cnt@rev.pol.rules.uiv.p1.old-decision.m2.score\endcsname{1}
\expandafter\def\csname pct@rev.pol.rules.uiv.p1.old-decision.m2.score\endcsname{83.3}
\expandafter\def\csname ppt@rev.pol.rules.uiv.p1.old-decision.m2.score\endcsname{$+$83.3}
\expandafter\def\csname val@rev.pol.rules.uiv.p1.oracle-on-T.keep\endcsname{0.000}
\expandafter\def\csname cnt@rev.pol.rules.uiv.p1.oracle-on-T.keep\endcsname{0}
\expandafter\def\csname pct@rev.pol.rules.uiv.p1.oracle-on-T.keep\endcsname{0.0}
\expandafter\def\csname ppt@rev.pol.rules.uiv.p1.oracle-on-T.keep\endcsname{0.0}
\expandafter\def\csname val@rev.pol.rules.uiv.p1.oracle-on-T.req\endcsname{1.000}
\expandafter\def\csname cnt@rev.pol.rules.uiv.p1.oracle-on-T.req\endcsname{1}
\expandafter\def\csname pct@rev.pol.rules.uiv.p1.oracle-on-T.req\endcsname{100.0}
\expandafter\def\csname ppt@rev.pol.rules.uiv.p1.oracle-on-T.req\endcsname{$+$100.0}
\expandafter\def\csname val@rev.pol.rules.uiv.p1.oracle-on-T.score\endcsname{0.833}
\expandafter\def\csname cnt@rev.pol.rules.uiv.p1.oracle-on-T.score\endcsname{1}
\expandafter\def\csname pct@rev.pol.rules.uiv.p1.oracle-on-T.score\endcsname{83.3}
\expandafter\def\csname ppt@rev.pol.rules.uiv.p1.oracle-on-T.score\endcsname{$+$83.3}
\expandafter\def\csname val@rev.pol.rules.uiv.p1.r-s.center-d1\endcsname{2.000}
\expandafter\def\csname cnt@rev.pol.rules.uiv.p1.r-s.center-d1\endcsname{2}
\expandafter\def\csname pct@rev.pol.rules.uiv.p1.r-s.center-d1\endcsname{200.0}
\expandafter\def\csname ppt@rev.pol.rules.uiv.p1.r-s.center-d1\endcsname{$+$200.0}
\expandafter\def\csname val@rev.pol.rules.uiv.p1.r-s.context-d1\endcsname{2.000}
\expandafter\def\csname cnt@rev.pol.rules.uiv.p1.r-s.context-d1\endcsname{2}
\expandafter\def\csname pct@rev.pol.rules.uiv.p1.r-s.context-d1\endcsname{200.0}
\expandafter\def\csname ppt@rev.pol.rules.uiv.p1.r-s.context-d1\endcsname{$+$200.0}
\expandafter\def\csname val@rev.pol.rules.uiv.p1.r-s.dual-d2\endcsname{3.031}
\expandafter\def\csname cnt@rev.pol.rules.uiv.p1.r-s.dual-d2\endcsname{3}
\expandafter\def\csname pct@rev.pol.rules.uiv.p1.r-s.dual-d2\endcsname{303.1}
\expandafter\def\csname ppt@rev.pol.rules.uiv.p1.r-s.dual-d2\endcsname{$+$303.1}
\expandafter\def\csname val@rev.pol.rules.uiv.p1.r-s.grid2-d3\endcsname{3.812}
\expandafter\def\csname cnt@rev.pol.rules.uiv.p1.r-s.grid2-d3\endcsname{4}
\expandafter\def\csname pct@rev.pol.rules.uiv.p1.r-s.grid2-d3\endcsname{381.2}
\expandafter\def\csname ppt@rev.pol.rules.uiv.p1.r-s.grid2-d3\endcsname{$+$381.2}
\expandafter\def\csname val@rev.pol.rules.uiv.p1.rcr-stratified@0.25.frozen.keep\endcsname{0.379}
\expandafter\def\csname cnt@rev.pol.rules.uiv.p1.rcr-stratified@0.25.frozen.keep\endcsname{0}
\expandafter\def\csname pct@rev.pol.rules.uiv.p1.rcr-stratified@0.25.frozen.keep\endcsname{37.9}
\expandafter\def\csname ppt@rev.pol.rules.uiv.p1.rcr-stratified@0.25.frozen.keep\endcsname{$+$37.9}
\expandafter\def\csname val@rev.pol.rules.uiv.p1.rcr-stratified@0.25.frozen.req\endcsname{1.626}
\expandafter\def\csname cnt@rev.pol.rules.uiv.p1.rcr-stratified@0.25.frozen.req\endcsname{2}
\expandafter\def\csname pct@rev.pol.rules.uiv.p1.rcr-stratified@0.25.frozen.req\endcsname{162.6}
\expandafter\def\csname ppt@rev.pol.rules.uiv.p1.rcr-stratified@0.25.frozen.req\endcsname{$+$162.6}
\expandafter\def\csname val@rev.pol.rules.uiv.p1.rcr-stratified@0.25.frozen.score\endcsname{0.805}
\expandafter\def\csname cnt@rev.pol.rules.uiv.p1.rcr-stratified@0.25.frozen.score\endcsname{1}
\expandafter\def\csname pct@rev.pol.rules.uiv.p1.rcr-stratified@0.25.frozen.score\endcsname{80.5}
\expandafter\def\csname ppt@rev.pol.rules.uiv.p1.rcr-stratified@0.25.frozen.score\endcsname{$+$80.5}
\expandafter\def\csname val@rev.pol.rules.uiv.p1.rcr-stratified@0.25.lam0.5.keep\endcsname{0.340}
\expandafter\def\csname cnt@rev.pol.rules.uiv.p1.rcr-stratified@0.25.lam0.5.keep\endcsname{0}
\expandafter\def\csname pct@rev.pol.rules.uiv.p1.rcr-stratified@0.25.lam0.5.keep\endcsname{34.0}
\expandafter\def\csname ppt@rev.pol.rules.uiv.p1.rcr-stratified@0.25.lam0.5.keep\endcsname{$+$34.0}
\expandafter\def\csname val@rev.pol.rules.uiv.p1.rcr-stratified@0.25.lam0.5.req\endcsname{1.538}
\expandafter\def\csname cnt@rev.pol.rules.uiv.p1.rcr-stratified@0.25.lam0.5.req\endcsname{2}
\expandafter\def\csname pct@rev.pol.rules.uiv.p1.rcr-stratified@0.25.lam0.5.req\endcsname{153.8}
\expandafter\def\csname ppt@rev.pol.rules.uiv.p1.rcr-stratified@0.25.lam0.5.req\endcsname{$+$153.8}
\expandafter\def\csname val@rev.pol.rules.uiv.p1.rcr-stratified@0.25.lam0.5.score\endcsname{0.808}
\expandafter\def\csname cnt@rev.pol.rules.uiv.p1.rcr-stratified@0.25.lam0.5.score\endcsname{1}
\expandafter\def\csname pct@rev.pol.rules.uiv.p1.rcr-stratified@0.25.lam0.5.score\endcsname{80.8}
\expandafter\def\csname ppt@rev.pol.rules.uiv.p1.rcr-stratified@0.25.lam0.5.score\endcsname{$+$80.8}
\expandafter\def\csname val@rev.pol.rules.uiv.p1.rcr-stratified@0.25.lam1.keep\endcsname{0.290}
\expandafter\def\csname cnt@rev.pol.rules.uiv.p1.rcr-stratified@0.25.lam1.keep\endcsname{0}
\expandafter\def\csname pct@rev.pol.rules.uiv.p1.rcr-stratified@0.25.lam1.keep\endcsname{29.0}
\expandafter\def\csname ppt@rev.pol.rules.uiv.p1.rcr-stratified@0.25.lam1.keep\endcsname{$+$29.0}
\expandafter\def\csname val@rev.pol.rules.uiv.p1.rcr-stratified@0.25.lam1.req\endcsname{1.446}
\expandafter\def\csname cnt@rev.pol.rules.uiv.p1.rcr-stratified@0.25.lam1.req\endcsname{1}
\expandafter\def\csname pct@rev.pol.rules.uiv.p1.rcr-stratified@0.25.lam1.req\endcsname{144.6}
\expandafter\def\csname ppt@rev.pol.rules.uiv.p1.rcr-stratified@0.25.lam1.req\endcsname{$+$144.6}
\expandafter\def\csname val@rev.pol.rules.uiv.p1.rcr-stratified@0.25.lam1.score\endcsname{0.812}
\expandafter\def\csname cnt@rev.pol.rules.uiv.p1.rcr-stratified@0.25.lam1.score\endcsname{1}
\expandafter\def\csname pct@rev.pol.rules.uiv.p1.rcr-stratified@0.25.lam1.score\endcsname{81.2}
\expandafter\def\csname ppt@rev.pol.rules.uiv.p1.rcr-stratified@0.25.lam1.score\endcsname{$+$81.2}
\expandafter\def\csname val@rev.pol.rules.uiv.p1.rcr-stratified@0.25.m0.5.keep\endcsname{0.358}
\expandafter\def\csname cnt@rev.pol.rules.uiv.p1.rcr-stratified@0.25.m0.5.keep\endcsname{0}
\expandafter\def\csname pct@rev.pol.rules.uiv.p1.rcr-stratified@0.25.m0.5.keep\endcsname{35.9}
\expandafter\def\csname ppt@rev.pol.rules.uiv.p1.rcr-stratified@0.25.m0.5.keep\endcsname{$+$35.9}
\expandafter\def\csname val@rev.pol.rules.uiv.p1.rcr-stratified@0.25.m0.5.req\endcsname{1.586}
\expandafter\def\csname cnt@rev.pol.rules.uiv.p1.rcr-stratified@0.25.m0.5.req\endcsname{2}
\expandafter\def\csname pct@rev.pol.rules.uiv.p1.rcr-stratified@0.25.m0.5.req\endcsname{158.6}
\expandafter\def\csname ppt@rev.pol.rules.uiv.p1.rcr-stratified@0.25.m0.5.req\endcsname{$+$158.6}
\expandafter\def\csname val@rev.pol.rules.uiv.p1.rcr-stratified@0.25.m0.5.score\endcsname{0.807}
\expandafter\def\csname cnt@rev.pol.rules.uiv.p1.rcr-stratified@0.25.m0.5.score\endcsname{1}
\expandafter\def\csname pct@rev.pol.rules.uiv.p1.rcr-stratified@0.25.m0.5.score\endcsname{80.7}
\expandafter\def\csname ppt@rev.pol.rules.uiv.p1.rcr-stratified@0.25.m0.5.score\endcsname{$+$80.7}
\expandafter\def\csname val@rev.pol.rules.uiv.p1.rcr-stratified@0.25.m1.keep\endcsname{0.343}
\expandafter\def\csname cnt@rev.pol.rules.uiv.p1.rcr-stratified@0.25.m1.keep\endcsname{0}
\expandafter\def\csname pct@rev.pol.rules.uiv.p1.rcr-stratified@0.25.m1.keep\endcsname{34.3}
\expandafter\def\csname ppt@rev.pol.rules.uiv.p1.rcr-stratified@0.25.m1.keep\endcsname{$+$34.3}
\expandafter\def\csname val@rev.pol.rules.uiv.p1.rcr-stratified@0.25.m1.req\endcsname{1.569}
\expandafter\def\csname cnt@rev.pol.rules.uiv.p1.rcr-stratified@0.25.m1.req\endcsname{2}
\expandafter\def\csname pct@rev.pol.rules.uiv.p1.rcr-stratified@0.25.m1.req\endcsname{156.9}
\expandafter\def\csname ppt@rev.pol.rules.uiv.p1.rcr-stratified@0.25.m1.req\endcsname{$+$156.9}
\expandafter\def\csname val@rev.pol.rules.uiv.p1.rcr-stratified@0.25.m1.score\endcsname{0.807}
\expandafter\def\csname cnt@rev.pol.rules.uiv.p1.rcr-stratified@0.25.m1.score\endcsname{1}
\expandafter\def\csname pct@rev.pol.rules.uiv.p1.rcr-stratified@0.25.m1.score\endcsname{80.7}
\expandafter\def\csname ppt@rev.pol.rules.uiv.p1.rcr-stratified@0.25.m1.score\endcsname{$+$80.7}
\expandafter\def\csname val@rev.pol.rules.uiv.p1.rcr-stratified@0.25.m2.keep\endcsname{0.245}
\expandafter\def\csname cnt@rev.pol.rules.uiv.p1.rcr-stratified@0.25.m2.keep\endcsname{0}
\expandafter\def\csname pct@rev.pol.rules.uiv.p1.rcr-stratified@0.25.m2.keep\endcsname{24.5}
\expandafter\def\csname ppt@rev.pol.rules.uiv.p1.rcr-stratified@0.25.m2.keep\endcsname{$+$24.5}
\expandafter\def\csname val@rev.pol.rules.uiv.p1.rcr-stratified@0.25.m2.req\endcsname{1.417}
\expandafter\def\csname cnt@rev.pol.rules.uiv.p1.rcr-stratified@0.25.m2.req\endcsname{1}
\expandafter\def\csname pct@rev.pol.rules.uiv.p1.rcr-stratified@0.25.m2.req\endcsname{141.7}
\expandafter\def\csname ppt@rev.pol.rules.uiv.p1.rcr-stratified@0.25.m2.req\endcsname{$+$141.7}
\expandafter\def\csname val@rev.pol.rules.uiv.p1.rcr-stratified@0.25.m2.score\endcsname{0.814}
\expandafter\def\csname cnt@rev.pol.rules.uiv.p1.rcr-stratified@0.25.m2.score\endcsname{1}
\expandafter\def\csname pct@rev.pol.rules.uiv.p1.rcr-stratified@0.25.m2.score\endcsname{81.4}
\expandafter\def\csname ppt@rev.pol.rules.uiv.p1.rcr-stratified@0.25.m2.score\endcsname{$+$81.4}
\expandafter\def\csname val@rev.pol.rules.uiv.p1.replay-only.frozen.keep\endcsname{0.000}
\expandafter\def\csname cnt@rev.pol.rules.uiv.p1.replay-only.frozen.keep\endcsname{0}
\expandafter\def\csname pct@rev.pol.rules.uiv.p1.replay-only.frozen.keep\endcsname{0.0}
\expandafter\def\csname ppt@rev.pol.rules.uiv.p1.replay-only.frozen.keep\endcsname{0.0}
\expandafter\def\csname val@rev.pol.rules.uiv.p1.replay-only.frozen.req\endcsname{1.000}
\expandafter\def\csname cnt@rev.pol.rules.uiv.p1.replay-only.frozen.req\endcsname{1}
\expandafter\def\csname pct@rev.pol.rules.uiv.p1.replay-only.frozen.req\endcsname{100.0}
\expandafter\def\csname ppt@rev.pol.rules.uiv.p1.replay-only.frozen.req\endcsname{$+$100.0}
\expandafter\def\csname val@rev.pol.rules.uiv.p1.replay-only.frozen.score\endcsname{0.833}
\expandafter\def\csname cnt@rev.pol.rules.uiv.p1.replay-only.frozen.score\endcsname{1}
\expandafter\def\csname pct@rev.pol.rules.uiv.p1.replay-only.frozen.score\endcsname{83.3}
\expandafter\def\csname ppt@rev.pol.rules.uiv.p1.replay-only.frozen.score\endcsname{$+$83.3}
\expandafter\def\csname val@rev.pol.rules.uiv.p1.replay-only.lam0.5.keep\endcsname{0.000}
\expandafter\def\csname cnt@rev.pol.rules.uiv.p1.replay-only.lam0.5.keep\endcsname{0}
\expandafter\def\csname pct@rev.pol.rules.uiv.p1.replay-only.lam0.5.keep\endcsname{0.0}
\expandafter\def\csname ppt@rev.pol.rules.uiv.p1.replay-only.lam0.5.keep\endcsname{0.0}
\expandafter\def\csname val@rev.pol.rules.uiv.p1.replay-only.lam0.5.req\endcsname{1.000}
\expandafter\def\csname cnt@rev.pol.rules.uiv.p1.replay-only.lam0.5.req\endcsname{1}
\expandafter\def\csname pct@rev.pol.rules.uiv.p1.replay-only.lam0.5.req\endcsname{100.0}
\expandafter\def\csname ppt@rev.pol.rules.uiv.p1.replay-only.lam0.5.req\endcsname{$+$100.0}
\expandafter\def\csname val@rev.pol.rules.uiv.p1.replay-only.lam0.5.score\endcsname{0.833}
\expandafter\def\csname cnt@rev.pol.rules.uiv.p1.replay-only.lam0.5.score\endcsname{1}
\expandafter\def\csname pct@rev.pol.rules.uiv.p1.replay-only.lam0.5.score\endcsname{83.3}
\expandafter\def\csname ppt@rev.pol.rules.uiv.p1.replay-only.lam0.5.score\endcsname{$+$83.3}
\expandafter\def\csname val@rev.pol.rules.uiv.p1.replay-only.lam1.keep\endcsname{0.000}
\expandafter\def\csname cnt@rev.pol.rules.uiv.p1.replay-only.lam1.keep\endcsname{0}
\expandafter\def\csname pct@rev.pol.rules.uiv.p1.replay-only.lam1.keep\endcsname{0.0}
\expandafter\def\csname ppt@rev.pol.rules.uiv.p1.replay-only.lam1.keep\endcsname{0.0}
\expandafter\def\csname val@rev.pol.rules.uiv.p1.replay-only.lam1.req\endcsname{1.000}
\expandafter\def\csname cnt@rev.pol.rules.uiv.p1.replay-only.lam1.req\endcsname{1}
\expandafter\def\csname pct@rev.pol.rules.uiv.p1.replay-only.lam1.req\endcsname{100.0}
\expandafter\def\csname ppt@rev.pol.rules.uiv.p1.replay-only.lam1.req\endcsname{$+$100.0}
\expandafter\def\csname val@rev.pol.rules.uiv.p1.replay-only.lam1.score\endcsname{0.833}
\expandafter\def\csname cnt@rev.pol.rules.uiv.p1.replay-only.lam1.score\endcsname{1}
\expandafter\def\csname pct@rev.pol.rules.uiv.p1.replay-only.lam1.score\endcsname{83.3}
\expandafter\def\csname ppt@rev.pol.rules.uiv.p1.replay-only.lam1.score\endcsname{$+$83.3}
\expandafter\def\csname val@rev.pol.rules.uiv.p1.replay-only.m0.5.keep\endcsname{0.000}
\expandafter\def\csname cnt@rev.pol.rules.uiv.p1.replay-only.m0.5.keep\endcsname{0}
\expandafter\def\csname pct@rev.pol.rules.uiv.p1.replay-only.m0.5.keep\endcsname{0.0}
\expandafter\def\csname ppt@rev.pol.rules.uiv.p1.replay-only.m0.5.keep\endcsname{0.0}
\expandafter\def\csname val@rev.pol.rules.uiv.p1.replay-only.m0.5.req\endcsname{1.000}
\expandafter\def\csname cnt@rev.pol.rules.uiv.p1.replay-only.m0.5.req\endcsname{1}
\expandafter\def\csname pct@rev.pol.rules.uiv.p1.replay-only.m0.5.req\endcsname{100.0}
\expandafter\def\csname ppt@rev.pol.rules.uiv.p1.replay-only.m0.5.req\endcsname{$+$100.0}
\expandafter\def\csname val@rev.pol.rules.uiv.p1.replay-only.m0.5.score\endcsname{0.833}
\expandafter\def\csname cnt@rev.pol.rules.uiv.p1.replay-only.m0.5.score\endcsname{1}
\expandafter\def\csname pct@rev.pol.rules.uiv.p1.replay-only.m0.5.score\endcsname{83.3}
\expandafter\def\csname ppt@rev.pol.rules.uiv.p1.replay-only.m0.5.score\endcsname{$+$83.3}
\expandafter\def\csname val@rev.pol.rules.uiv.p1.replay-only.m1.keep\endcsname{0.000}
\expandafter\def\csname cnt@rev.pol.rules.uiv.p1.replay-only.m1.keep\endcsname{0}
\expandafter\def\csname pct@rev.pol.rules.uiv.p1.replay-only.m1.keep\endcsname{0.0}
\expandafter\def\csname ppt@rev.pol.rules.uiv.p1.replay-only.m1.keep\endcsname{0.0}
\expandafter\def\csname val@rev.pol.rules.uiv.p1.replay-only.m1.req\endcsname{1.000}
\expandafter\def\csname cnt@rev.pol.rules.uiv.p1.replay-only.m1.req\endcsname{1}
\expandafter\def\csname pct@rev.pol.rules.uiv.p1.replay-only.m1.req\endcsname{100.0}
\expandafter\def\csname ppt@rev.pol.rules.uiv.p1.replay-only.m1.req\endcsname{$+$100.0}
\expandafter\def\csname val@rev.pol.rules.uiv.p1.replay-only.m1.score\endcsname{0.833}
\expandafter\def\csname cnt@rev.pol.rules.uiv.p1.replay-only.m1.score\endcsname{1}
\expandafter\def\csname pct@rev.pol.rules.uiv.p1.replay-only.m1.score\endcsname{83.3}
\expandafter\def\csname ppt@rev.pol.rules.uiv.p1.replay-only.m1.score\endcsname{$+$83.3}
\expandafter\def\csname val@rev.pol.rules.uiv.p1.replay-only.m2.keep\endcsname{0.000}
\expandafter\def\csname cnt@rev.pol.rules.uiv.p1.replay-only.m2.keep\endcsname{0}
\expandafter\def\csname pct@rev.pol.rules.uiv.p1.replay-only.m2.keep\endcsname{0.0}
\expandafter\def\csname ppt@rev.pol.rules.uiv.p1.replay-only.m2.keep\endcsname{0.0}
\expandafter\def\csname val@rev.pol.rules.uiv.p1.replay-only.m2.req\endcsname{1.000}
\expandafter\def\csname cnt@rev.pol.rules.uiv.p1.replay-only.m2.req\endcsname{1}
\expandafter\def\csname pct@rev.pol.rules.uiv.p1.replay-only.m2.req\endcsname{100.0}
\expandafter\def\csname ppt@rev.pol.rules.uiv.p1.replay-only.m2.req\endcsname{$+$100.0}
\expandafter\def\csname val@rev.pol.rules.uiv.p1.replay-only.m2.score\endcsname{0.833}
\expandafter\def\csname cnt@rev.pol.rules.uiv.p1.replay-only.m2.score\endcsname{1}
\expandafter\def\csname pct@rev.pol.rules.uiv.p1.replay-only.m2.score\endcsname{83.3}
\expandafter\def\csname ppt@rev.pol.rules.uiv.p1.replay-only.m2.score\endcsname{$+$83.3}
\expandafter\def\csname val@rev.pol.rules.uiv.p1.replay-ppi-uniform@0.25.frozen.keep\endcsname{0.435}
\expandafter\def\csname cnt@rev.pol.rules.uiv.p1.replay-ppi-uniform@0.25.frozen.keep\endcsname{0}
\expandafter\def\csname pct@rev.pol.rules.uiv.p1.replay-ppi-uniform@0.25.frozen.keep\endcsname{43.5}
\expandafter\def\csname ppt@rev.pol.rules.uiv.p1.replay-ppi-uniform@0.25.frozen.keep\endcsname{$+$43.5}
\expandafter\def\csname val@rev.pol.rules.uiv.p1.replay-ppi-uniform@0.25.frozen.req\endcsname{1.733}
\expandafter\def\csname cnt@rev.pol.rules.uiv.p1.replay-ppi-uniform@0.25.frozen.req\endcsname{2}
\expandafter\def\csname pct@rev.pol.rules.uiv.p1.replay-ppi-uniform@0.25.frozen.req\endcsname{173.3}
\expandafter\def\csname ppt@rev.pol.rules.uiv.p1.replay-ppi-uniform@0.25.frozen.req\endcsname{$+$173.3}
\expandafter\def\csname val@rev.pol.rules.uiv.p1.replay-ppi-uniform@0.25.frozen.score\endcsname{0.800}
\expandafter\def\csname cnt@rev.pol.rules.uiv.p1.replay-ppi-uniform@0.25.frozen.score\endcsname{1}
\expandafter\def\csname pct@rev.pol.rules.uiv.p1.replay-ppi-uniform@0.25.frozen.score\endcsname{80.0}
\expandafter\def\csname ppt@rev.pol.rules.uiv.p1.replay-ppi-uniform@0.25.frozen.score\endcsname{$+$80.0}
\expandafter\def\csname val@rev.pol.rules.uiv.p1.replay-ppi-uniform@0.25.lam0.5.keep\endcsname{0.388}
\expandafter\def\csname cnt@rev.pol.rules.uiv.p1.replay-ppi-uniform@0.25.lam0.5.keep\endcsname{0}
\expandafter\def\csname pct@rev.pol.rules.uiv.p1.replay-ppi-uniform@0.25.lam0.5.keep\endcsname{38.8}
\expandafter\def\csname ppt@rev.pol.rules.uiv.p1.replay-ppi-uniform@0.25.lam0.5.keep\endcsname{$+$38.8}
\expandafter\def\csname val@rev.pol.rules.uiv.p1.replay-ppi-uniform@0.25.lam0.5.req\endcsname{1.609}
\expandafter\def\csname cnt@rev.pol.rules.uiv.p1.replay-ppi-uniform@0.25.lam0.5.req\endcsname{2}
\expandafter\def\csname pct@rev.pol.rules.uiv.p1.replay-ppi-uniform@0.25.lam0.5.req\endcsname{160.9}
\expandafter\def\csname ppt@rev.pol.rules.uiv.p1.replay-ppi-uniform@0.25.lam0.5.req\endcsname{$+$160.9}
\expandafter\def\csname val@rev.pol.rules.uiv.p1.replay-ppi-uniform@0.25.lam0.5.score\endcsname{0.804}
\expandafter\def\csname cnt@rev.pol.rules.uiv.p1.replay-ppi-uniform@0.25.lam0.5.score\endcsname{1}
\expandafter\def\csname pct@rev.pol.rules.uiv.p1.replay-ppi-uniform@0.25.lam0.5.score\endcsname{80.4}
\expandafter\def\csname ppt@rev.pol.rules.uiv.p1.replay-ppi-uniform@0.25.lam0.5.score\endcsname{$+$80.4}
\expandafter\def\csname val@rev.pol.rules.uiv.p1.replay-ppi-uniform@0.25.lam1.keep\endcsname{0.322}
\expandafter\def\csname cnt@rev.pol.rules.uiv.p1.replay-ppi-uniform@0.25.lam1.keep\endcsname{0}
\expandafter\def\csname pct@rev.pol.rules.uiv.p1.replay-ppi-uniform@0.25.lam1.keep\endcsname{32.2}
\expandafter\def\csname ppt@rev.pol.rules.uiv.p1.replay-ppi-uniform@0.25.lam1.keep\endcsname{$+$32.2}
\expandafter\def\csname val@rev.pol.rules.uiv.p1.replay-ppi-uniform@0.25.lam1.req\endcsname{1.474}
\expandafter\def\csname cnt@rev.pol.rules.uiv.p1.replay-ppi-uniform@0.25.lam1.req\endcsname{1}
\expandafter\def\csname pct@rev.pol.rules.uiv.p1.replay-ppi-uniform@0.25.lam1.req\endcsname{147.4}
\expandafter\def\csname ppt@rev.pol.rules.uiv.p1.replay-ppi-uniform@0.25.lam1.req\endcsname{$+$147.4}
\expandafter\def\csname val@rev.pol.rules.uiv.p1.replay-ppi-uniform@0.25.lam1.score\endcsname{0.811}
\expandafter\def\csname cnt@rev.pol.rules.uiv.p1.replay-ppi-uniform@0.25.lam1.score\endcsname{1}
\expandafter\def\csname pct@rev.pol.rules.uiv.p1.replay-ppi-uniform@0.25.lam1.score\endcsname{81.1}
\expandafter\def\csname ppt@rev.pol.rules.uiv.p1.replay-ppi-uniform@0.25.lam1.score\endcsname{$+$81.1}
\expandafter\def\csname val@rev.pol.rules.uiv.p1.replay-ppi-uniform@0.25.m0.5.keep\endcsname{0.435}
\expandafter\def\csname cnt@rev.pol.rules.uiv.p1.replay-ppi-uniform@0.25.m0.5.keep\endcsname{0}
\expandafter\def\csname pct@rev.pol.rules.uiv.p1.replay-ppi-uniform@0.25.m0.5.keep\endcsname{43.5}
\expandafter\def\csname ppt@rev.pol.rules.uiv.p1.replay-ppi-uniform@0.25.m0.5.keep\endcsname{$+$43.5}
\expandafter\def\csname val@rev.pol.rules.uiv.p1.replay-ppi-uniform@0.25.m0.5.req\endcsname{1.733}
\expandafter\def\csname cnt@rev.pol.rules.uiv.p1.replay-ppi-uniform@0.25.m0.5.req\endcsname{2}
\expandafter\def\csname pct@rev.pol.rules.uiv.p1.replay-ppi-uniform@0.25.m0.5.req\endcsname{173.3}
\expandafter\def\csname ppt@rev.pol.rules.uiv.p1.replay-ppi-uniform@0.25.m0.5.req\endcsname{$+$173.3}
\expandafter\def\csname val@rev.pol.rules.uiv.p1.replay-ppi-uniform@0.25.m0.5.score\endcsname{0.800}
\expandafter\def\csname cnt@rev.pol.rules.uiv.p1.replay-ppi-uniform@0.25.m0.5.score\endcsname{1}
\expandafter\def\csname pct@rev.pol.rules.uiv.p1.replay-ppi-uniform@0.25.m0.5.score\endcsname{80.0}
\expandafter\def\csname ppt@rev.pol.rules.uiv.p1.replay-ppi-uniform@0.25.m0.5.score\endcsname{$+$80.0}
\expandafter\def\csname val@rev.pol.rules.uiv.p1.replay-ppi-uniform@0.25.m1.keep\endcsname{0.435}
\expandafter\def\csname cnt@rev.pol.rules.uiv.p1.replay-ppi-uniform@0.25.m1.keep\endcsname{0}
\expandafter\def\csname pct@rev.pol.rules.uiv.p1.replay-ppi-uniform@0.25.m1.keep\endcsname{43.5}
\expandafter\def\csname ppt@rev.pol.rules.uiv.p1.replay-ppi-uniform@0.25.m1.keep\endcsname{$+$43.5}
\expandafter\def\csname val@rev.pol.rules.uiv.p1.replay-ppi-uniform@0.25.m1.req\endcsname{1.733}
\expandafter\def\csname cnt@rev.pol.rules.uiv.p1.replay-ppi-uniform@0.25.m1.req\endcsname{2}
\expandafter\def\csname pct@rev.pol.rules.uiv.p1.replay-ppi-uniform@0.25.m1.req\endcsname{173.3}
\expandafter\def\csname ppt@rev.pol.rules.uiv.p1.replay-ppi-uniform@0.25.m1.req\endcsname{$+$173.3}
\expandafter\def\csname val@rev.pol.rules.uiv.p1.replay-ppi-uniform@0.25.m1.score\endcsname{0.800}
\expandafter\def\csname cnt@rev.pol.rules.uiv.p1.replay-ppi-uniform@0.25.m1.score\endcsname{1}
\expandafter\def\csname pct@rev.pol.rules.uiv.p1.replay-ppi-uniform@0.25.m1.score\endcsname{80.0}
\expandafter\def\csname ppt@rev.pol.rules.uiv.p1.replay-ppi-uniform@0.25.m1.score\endcsname{$+$80.0}
\expandafter\def\csname val@rev.pol.rules.uiv.p1.replay-ppi-uniform@0.25.m2.keep\endcsname{0.322}
\expandafter\def\csname cnt@rev.pol.rules.uiv.p1.replay-ppi-uniform@0.25.m2.keep\endcsname{0}
\expandafter\def\csname pct@rev.pol.rules.uiv.p1.replay-ppi-uniform@0.25.m2.keep\endcsname{32.2}
\expandafter\def\csname ppt@rev.pol.rules.uiv.p1.replay-ppi-uniform@0.25.m2.keep\endcsname{$+$32.2}
\expandafter\def\csname val@rev.pol.rules.uiv.p1.replay-ppi-uniform@0.25.m2.req\endcsname{1.474}
\expandafter\def\csname cnt@rev.pol.rules.uiv.p1.replay-ppi-uniform@0.25.m2.req\endcsname{1}
\expandafter\def\csname pct@rev.pol.rules.uiv.p1.replay-ppi-uniform@0.25.m2.req\endcsname{147.4}
\expandafter\def\csname ppt@rev.pol.rules.uiv.p1.replay-ppi-uniform@0.25.m2.req\endcsname{$+$147.4}
\expandafter\def\csname val@rev.pol.rules.uiv.p1.replay-ppi-uniform@0.25.m2.score\endcsname{0.811}
\expandafter\def\csname cnt@rev.pol.rules.uiv.p1.replay-ppi-uniform@0.25.m2.score\endcsname{1}
\expandafter\def\csname pct@rev.pol.rules.uiv.p1.replay-ppi-uniform@0.25.m2.score\endcsname{81.1}
\expandafter\def\csname ppt@rev.pol.rules.uiv.p1.replay-ppi-uniform@0.25.m2.score\endcsname{$+$81.1}
\expandafter\def\csname val@rev.pol.rules.uiv.p1.suspend-all.keep\endcsname{0.000}
\expandafter\def\csname cnt@rev.pol.rules.uiv.p1.suspend-all.keep\endcsname{0}
\expandafter\def\csname pct@rev.pol.rules.uiv.p1.suspend-all.keep\endcsname{0.0}
\expandafter\def\csname ppt@rev.pol.rules.uiv.p1.suspend-all.keep\endcsname{0.0}
\expandafter\def\csname val@rev.pol.rules.uiv.p1.suspend-all.req\endcsname{1.000}
\expandafter\def\csname cnt@rev.pol.rules.uiv.p1.suspend-all.req\endcsname{1}
\expandafter\def\csname pct@rev.pol.rules.uiv.p1.suspend-all.req\endcsname{100.0}
\expandafter\def\csname ppt@rev.pol.rules.uiv.p1.suspend-all.req\endcsname{$+$100.0}
\expandafter\def\csname val@rev.pol.rules.uiv.p1.suspend-all.score\endcsname{0.833}
\expandafter\def\csname cnt@rev.pol.rules.uiv.p1.suspend-all.score\endcsname{1}
\expandafter\def\csname pct@rev.pol.rules.uiv.p1.suspend-all.score\endcsname{83.3}
\expandafter\def\csname ppt@rev.pol.rules.uiv.p1.suspend-all.score\endcsname{$+$83.3}
\expandafter\def\csname val@rev.pol.rules.uiv.p1.uniform-fresh@0.25.frozen.keep\endcsname{0.570}
\expandafter\def\csname cnt@rev.pol.rules.uiv.p1.uniform-fresh@0.25.frozen.keep\endcsname{1}
\expandafter\def\csname pct@rev.pol.rules.uiv.p1.uniform-fresh@0.25.frozen.keep\endcsname{57.0}
\expandafter\def\csname ppt@rev.pol.rules.uiv.p1.uniform-fresh@0.25.frozen.keep\endcsname{$+$57.0}
\expandafter\def\csname val@rev.pol.rules.uiv.p1.uniform-fresh@0.25.frozen.req\endcsname{1.913}
\expandafter\def\csname cnt@rev.pol.rules.uiv.p1.uniform-fresh@0.25.frozen.req\endcsname{2}
\expandafter\def\csname pct@rev.pol.rules.uiv.p1.uniform-fresh@0.25.frozen.req\endcsname{191.3}
\expandafter\def\csname ppt@rev.pol.rules.uiv.p1.uniform-fresh@0.25.frozen.req\endcsname{$+$191.3}
\expandafter\def\csname val@rev.pol.rules.uiv.p1.uniform-fresh@0.25.frozen.score\endcsname{0.792}
\expandafter\def\csname cnt@rev.pol.rules.uiv.p1.uniform-fresh@0.25.frozen.score\endcsname{1}
\expandafter\def\csname pct@rev.pol.rules.uiv.p1.uniform-fresh@0.25.frozen.score\endcsname{79.2}
\expandafter\def\csname ppt@rev.pol.rules.uiv.p1.uniform-fresh@0.25.frozen.score\endcsname{$+$79.2}
\expandafter\def\csname val@rev.pol.rules.uiv.p1.uniform-fresh@0.25.lam0.5.keep\endcsname{0.310}
\expandafter\def\csname cnt@rev.pol.rules.uiv.p1.uniform-fresh@0.25.lam0.5.keep\endcsname{0}
\expandafter\def\csname pct@rev.pol.rules.uiv.p1.uniform-fresh@0.25.lam0.5.keep\endcsname{31.0}
\expandafter\def\csname ppt@rev.pol.rules.uiv.p1.uniform-fresh@0.25.lam0.5.keep\endcsname{$+$31.0}
\expandafter\def\csname val@rev.pol.rules.uiv.p1.uniform-fresh@0.25.lam0.5.req\endcsname{1.485}
\expandafter\def\csname cnt@rev.pol.rules.uiv.p1.uniform-fresh@0.25.lam0.5.req\endcsname{1}
\expandafter\def\csname pct@rev.pol.rules.uiv.p1.uniform-fresh@0.25.lam0.5.req\endcsname{148.5}
\expandafter\def\csname ppt@rev.pol.rules.uiv.p1.uniform-fresh@0.25.lam0.5.req\endcsname{$+$148.5}
\expandafter\def\csname val@rev.pol.rules.uiv.p1.uniform-fresh@0.25.lam0.5.score\endcsname{0.811}
\expandafter\def\csname cnt@rev.pol.rules.uiv.p1.uniform-fresh@0.25.lam0.5.score\endcsname{1}
\expandafter\def\csname pct@rev.pol.rules.uiv.p1.uniform-fresh@0.25.lam0.5.score\endcsname{81.1}
\expandafter\def\csname ppt@rev.pol.rules.uiv.p1.uniform-fresh@0.25.lam0.5.score\endcsname{$+$81.1}
\expandafter\def\csname val@rev.pol.rules.uiv.p1.uniform-fresh@0.25.lam1.keep\endcsname{0.310}
\expandafter\def\csname cnt@rev.pol.rules.uiv.p1.uniform-fresh@0.25.lam1.keep\endcsname{0}
\expandafter\def\csname pct@rev.pol.rules.uiv.p1.uniform-fresh@0.25.lam1.keep\endcsname{31.0}
\expandafter\def\csname ppt@rev.pol.rules.uiv.p1.uniform-fresh@0.25.lam1.keep\endcsname{$+$31.0}
\expandafter\def\csname val@rev.pol.rules.uiv.p1.uniform-fresh@0.25.lam1.req\endcsname{1.485}
\expandafter\def\csname cnt@rev.pol.rules.uiv.p1.uniform-fresh@0.25.lam1.req\endcsname{1}
\expandafter\def\csname pct@rev.pol.rules.uiv.p1.uniform-fresh@0.25.lam1.req\endcsname{148.5}
\expandafter\def\csname ppt@rev.pol.rules.uiv.p1.uniform-fresh@0.25.lam1.req\endcsname{$+$148.5}
\expandafter\def\csname val@rev.pol.rules.uiv.p1.uniform-fresh@0.25.lam1.score\endcsname{0.811}
\expandafter\def\csname cnt@rev.pol.rules.uiv.p1.uniform-fresh@0.25.lam1.score\endcsname{1}
\expandafter\def\csname pct@rev.pol.rules.uiv.p1.uniform-fresh@0.25.lam1.score\endcsname{81.1}
\expandafter\def\csname ppt@rev.pol.rules.uiv.p1.uniform-fresh@0.25.lam1.score\endcsname{$+$81.1}
\expandafter\def\csname val@rev.pol.rules.uiv.p1.uniform-fresh@0.25.m0.5.keep\endcsname{0.310}
\expandafter\def\csname cnt@rev.pol.rules.uiv.p1.uniform-fresh@0.25.m0.5.keep\endcsname{0}
\expandafter\def\csname pct@rev.pol.rules.uiv.p1.uniform-fresh@0.25.m0.5.keep\endcsname{31.0}
\expandafter\def\csname ppt@rev.pol.rules.uiv.p1.uniform-fresh@0.25.m0.5.keep\endcsname{$+$31.0}
\expandafter\def\csname val@rev.pol.rules.uiv.p1.uniform-fresh@0.25.m0.5.req\endcsname{1.485}
\expandafter\def\csname cnt@rev.pol.rules.uiv.p1.uniform-fresh@0.25.m0.5.req\endcsname{1}
\expandafter\def\csname pct@rev.pol.rules.uiv.p1.uniform-fresh@0.25.m0.5.req\endcsname{148.5}
\expandafter\def\csname ppt@rev.pol.rules.uiv.p1.uniform-fresh@0.25.m0.5.req\endcsname{$+$148.5}
\expandafter\def\csname val@rev.pol.rules.uiv.p1.uniform-fresh@0.25.m0.5.score\endcsname{0.811}
\expandafter\def\csname cnt@rev.pol.rules.uiv.p1.uniform-fresh@0.25.m0.5.score\endcsname{1}
\expandafter\def\csname pct@rev.pol.rules.uiv.p1.uniform-fresh@0.25.m0.5.score\endcsname{81.1}
\expandafter\def\csname ppt@rev.pol.rules.uiv.p1.uniform-fresh@0.25.m0.5.score\endcsname{$+$81.1}
\expandafter\def\csname val@rev.pol.rules.uiv.p1.uniform-fresh@0.25.m1.keep\endcsname{0.310}
\expandafter\def\csname cnt@rev.pol.rules.uiv.p1.uniform-fresh@0.25.m1.keep\endcsname{0}
\expandafter\def\csname pct@rev.pol.rules.uiv.p1.uniform-fresh@0.25.m1.keep\endcsname{31.0}
\expandafter\def\csname ppt@rev.pol.rules.uiv.p1.uniform-fresh@0.25.m1.keep\endcsname{$+$31.0}
\expandafter\def\csname val@rev.pol.rules.uiv.p1.uniform-fresh@0.25.m1.req\endcsname{1.485}
\expandafter\def\csname cnt@rev.pol.rules.uiv.p1.uniform-fresh@0.25.m1.req\endcsname{1}
\expandafter\def\csname pct@rev.pol.rules.uiv.p1.uniform-fresh@0.25.m1.req\endcsname{148.5}
\expandafter\def\csname ppt@rev.pol.rules.uiv.p1.uniform-fresh@0.25.m1.req\endcsname{$+$148.5}
\expandafter\def\csname val@rev.pol.rules.uiv.p1.uniform-fresh@0.25.m1.score\endcsname{0.811}
\expandafter\def\csname cnt@rev.pol.rules.uiv.p1.uniform-fresh@0.25.m1.score\endcsname{1}
\expandafter\def\csname pct@rev.pol.rules.uiv.p1.uniform-fresh@0.25.m1.score\endcsname{81.1}
\expandafter\def\csname ppt@rev.pol.rules.uiv.p1.uniform-fresh@0.25.m1.score\endcsname{$+$81.1}
\expandafter\def\csname val@rev.pol.rules.uiv.p1.uniform-fresh@0.25.m2.keep\endcsname{0.310}
\expandafter\def\csname cnt@rev.pol.rules.uiv.p1.uniform-fresh@0.25.m2.keep\endcsname{0}
\expandafter\def\csname pct@rev.pol.rules.uiv.p1.uniform-fresh@0.25.m2.keep\endcsname{31.0}
\expandafter\def\csname ppt@rev.pol.rules.uiv.p1.uniform-fresh@0.25.m2.keep\endcsname{$+$31.0}
\expandafter\def\csname val@rev.pol.rules.uiv.p1.uniform-fresh@0.25.m2.req\endcsname{1.485}
\expandafter\def\csname cnt@rev.pol.rules.uiv.p1.uniform-fresh@0.25.m2.req\endcsname{1}
\expandafter\def\csname pct@rev.pol.rules.uiv.p1.uniform-fresh@0.25.m2.req\endcsname{148.5}
\expandafter\def\csname ppt@rev.pol.rules.uiv.p1.uniform-fresh@0.25.m2.req\endcsname{$+$148.5}
\expandafter\def\csname val@rev.pol.rules.uiv.p1.uniform-fresh@0.25.m2.score\endcsname{0.811}
\expandafter\def\csname cnt@rev.pol.rules.uiv.p1.uniform-fresh@0.25.m2.score\endcsname{1}
\expandafter\def\csname pct@rev.pol.rules.uiv.p1.uniform-fresh@0.25.m2.score\endcsname{81.1}
\expandafter\def\csname ppt@rev.pol.rules.uiv.p1.uniform-fresh@0.25.m2.score\endcsname{$+$81.1}
\expandafter\def\csname val@rev.pol.rules.uiv.p2.full-recert.frozen.keep\endcsname{0.750}
\expandafter\def\csname cnt@rev.pol.rules.uiv.p2.full-recert.frozen.keep\endcsname{1}
\expandafter\def\csname pct@rev.pol.rules.uiv.p2.full-recert.frozen.keep\endcsname{75.0}
\expandafter\def\csname ppt@rev.pol.rules.uiv.p2.full-recert.frozen.keep\endcsname{$+$75.0}
\expandafter\def\csname val@rev.pol.rules.uiv.p2.full-recert.frozen.req\endcsname{2.146}
\expandafter\def\csname cnt@rev.pol.rules.uiv.p2.full-recert.frozen.req\endcsname{2}
\expandafter\def\csname pct@rev.pol.rules.uiv.p2.full-recert.frozen.req\endcsname{214.6}
\expandafter\def\csname ppt@rev.pol.rules.uiv.p2.full-recert.frozen.req\endcsname{$+$214.6}
\expandafter\def\csname val@rev.pol.rules.uiv.p2.full-recert.frozen.score\endcsname{0.832}
\expandafter\def\csname cnt@rev.pol.rules.uiv.p2.full-recert.frozen.score\endcsname{1}
\expandafter\def\csname pct@rev.pol.rules.uiv.p2.full-recert.frozen.score\endcsname{83.2}
\expandafter\def\csname ppt@rev.pol.rules.uiv.p2.full-recert.frozen.score\endcsname{$+$83.2}
\expandafter\def\csname val@rev.pol.rules.uiv.p2.full-recert.lam0.5.keep\endcsname{0.500}
\expandafter\def\csname cnt@rev.pol.rules.uiv.p2.full-recert.lam0.5.keep\endcsname{0}
\expandafter\def\csname pct@rev.pol.rules.uiv.p2.full-recert.lam0.5.keep\endcsname{50.0}
\expandafter\def\csname ppt@rev.pol.rules.uiv.p2.full-recert.lam0.5.keep\endcsname{$+$50.0}
\expandafter\def\csname val@rev.pol.rules.uiv.p2.full-recert.lam0.5.req\endcsname{1.500}
\expandafter\def\csname cnt@rev.pol.rules.uiv.p2.full-recert.lam0.5.req\endcsname{2}
\expandafter\def\csname pct@rev.pol.rules.uiv.p2.full-recert.lam0.5.req\endcsname{150.0}
\expandafter\def\csname ppt@rev.pol.rules.uiv.p2.full-recert.lam0.5.req\endcsname{$+$150.0}
\expandafter\def\csname val@rev.pol.rules.uiv.p2.full-recert.lam0.5.score\endcsname{0.833}
\expandafter\def\csname cnt@rev.pol.rules.uiv.p2.full-recert.lam0.5.score\endcsname{1}
\expandafter\def\csname pct@rev.pol.rules.uiv.p2.full-recert.lam0.5.score\endcsname{83.3}
\expandafter\def\csname ppt@rev.pol.rules.uiv.p2.full-recert.lam0.5.score\endcsname{$+$83.3}
\expandafter\def\csname val@rev.pol.rules.uiv.p2.full-recert.lam1.keep\endcsname{0.500}
\expandafter\def\csname cnt@rev.pol.rules.uiv.p2.full-recert.lam1.keep\endcsname{0}
\expandafter\def\csname pct@rev.pol.rules.uiv.p2.full-recert.lam1.keep\endcsname{50.0}
\expandafter\def\csname ppt@rev.pol.rules.uiv.p2.full-recert.lam1.keep\endcsname{$+$50.0}
\expandafter\def\csname val@rev.pol.rules.uiv.p2.full-recert.lam1.req\endcsname{1.500}
\expandafter\def\csname cnt@rev.pol.rules.uiv.p2.full-recert.lam1.req\endcsname{2}
\expandafter\def\csname pct@rev.pol.rules.uiv.p2.full-recert.lam1.req\endcsname{150.0}
\expandafter\def\csname ppt@rev.pol.rules.uiv.p2.full-recert.lam1.req\endcsname{$+$150.0}
\expandafter\def\csname val@rev.pol.rules.uiv.p2.full-recert.lam1.score\endcsname{0.833}
\expandafter\def\csname cnt@rev.pol.rules.uiv.p2.full-recert.lam1.score\endcsname{1}
\expandafter\def\csname pct@rev.pol.rules.uiv.p2.full-recert.lam1.score\endcsname{83.3}
\expandafter\def\csname ppt@rev.pol.rules.uiv.p2.full-recert.lam1.score\endcsname{$+$83.3}
\expandafter\def\csname val@rev.pol.rules.uiv.p2.full-recert.m0.5.keep\endcsname{0.750}
\expandafter\def\csname cnt@rev.pol.rules.uiv.p2.full-recert.m0.5.keep\endcsname{1}
\expandafter\def\csname pct@rev.pol.rules.uiv.p2.full-recert.m0.5.keep\endcsname{75.0}
\expandafter\def\csname ppt@rev.pol.rules.uiv.p2.full-recert.m0.5.keep\endcsname{$+$75.0}
\expandafter\def\csname val@rev.pol.rules.uiv.p2.full-recert.m0.5.req\endcsname{2.146}
\expandafter\def\csname cnt@rev.pol.rules.uiv.p2.full-recert.m0.5.req\endcsname{2}
\expandafter\def\csname pct@rev.pol.rules.uiv.p2.full-recert.m0.5.req\endcsname{214.6}
\expandafter\def\csname ppt@rev.pol.rules.uiv.p2.full-recert.m0.5.req\endcsname{$+$214.6}
\expandafter\def\csname val@rev.pol.rules.uiv.p2.full-recert.m0.5.score\endcsname{0.832}
\expandafter\def\csname cnt@rev.pol.rules.uiv.p2.full-recert.m0.5.score\endcsname{1}
\expandafter\def\csname pct@rev.pol.rules.uiv.p2.full-recert.m0.5.score\endcsname{83.2}
\expandafter\def\csname ppt@rev.pol.rules.uiv.p2.full-recert.m0.5.score\endcsname{$+$83.2}
\expandafter\def\csname val@rev.pol.rules.uiv.p2.full-recert.m1.keep\endcsname{0.750}
\expandafter\def\csname cnt@rev.pol.rules.uiv.p2.full-recert.m1.keep\endcsname{1}
\expandafter\def\csname pct@rev.pol.rules.uiv.p2.full-recert.m1.keep\endcsname{75.0}
\expandafter\def\csname ppt@rev.pol.rules.uiv.p2.full-recert.m1.keep\endcsname{$+$75.0}
\expandafter\def\csname val@rev.pol.rules.uiv.p2.full-recert.m1.req\endcsname{2.146}
\expandafter\def\csname cnt@rev.pol.rules.uiv.p2.full-recert.m1.req\endcsname{2}
\expandafter\def\csname pct@rev.pol.rules.uiv.p2.full-recert.m1.req\endcsname{214.6}
\expandafter\def\csname ppt@rev.pol.rules.uiv.p2.full-recert.m1.req\endcsname{$+$214.6}
\expandafter\def\csname val@rev.pol.rules.uiv.p2.full-recert.m1.score\endcsname{0.832}
\expandafter\def\csname cnt@rev.pol.rules.uiv.p2.full-recert.m1.score\endcsname{1}
\expandafter\def\csname pct@rev.pol.rules.uiv.p2.full-recert.m1.score\endcsname{83.2}
\expandafter\def\csname ppt@rev.pol.rules.uiv.p2.full-recert.m1.score\endcsname{$+$83.2}
\expandafter\def\csname val@rev.pol.rules.uiv.p2.full-recert.m2.keep\endcsname{0.250}
\expandafter\def\csname cnt@rev.pol.rules.uiv.p2.full-recert.m2.keep\endcsname{0}
\expandafter\def\csname pct@rev.pol.rules.uiv.p2.full-recert.m2.keep\endcsname{25.0}
\expandafter\def\csname ppt@rev.pol.rules.uiv.p2.full-recert.m2.keep\endcsname{$+$25.0}
\expandafter\def\csname val@rev.pol.rules.uiv.p2.full-recert.m2.req\endcsname{1.250}
\expandafter\def\csname cnt@rev.pol.rules.uiv.p2.full-recert.m2.req\endcsname{1}
\expandafter\def\csname pct@rev.pol.rules.uiv.p2.full-recert.m2.req\endcsname{125.0}
\expandafter\def\csname ppt@rev.pol.rules.uiv.p2.full-recert.m2.req\endcsname{$+$125.0}
\expandafter\def\csname val@rev.pol.rules.uiv.p2.full-recert.m2.score\endcsname{0.833}
\expandafter\def\csname cnt@rev.pol.rules.uiv.p2.full-recert.m2.score\endcsname{1}
\expandafter\def\csname pct@rev.pol.rules.uiv.p2.full-recert.m2.score\endcsname{83.3}
\expandafter\def\csname ppt@rev.pol.rules.uiv.p2.full-recert.m2.score\endcsname{$+$83.3}
\expandafter\def\csname val@rev.pol.rules.uiv.p2.generic-ppi-source@0.25.frozen.keep\endcsname{0.544}
\expandafter\def\csname cnt@rev.pol.rules.uiv.p2.generic-ppi-source@0.25.frozen.keep\endcsname{1}
\expandafter\def\csname pct@rev.pol.rules.uiv.p2.generic-ppi-source@0.25.frozen.keep\endcsname{54.4}
\expandafter\def\csname ppt@rev.pol.rules.uiv.p2.generic-ppi-source@0.25.frozen.keep\endcsname{$+$54.4}
\expandafter\def\csname val@rev.pol.rules.uiv.p2.generic-ppi-source@0.25.frozen.req\endcsname{1.886}
\expandafter\def\csname cnt@rev.pol.rules.uiv.p2.generic-ppi-source@0.25.frozen.req\endcsname{2}
\expandafter\def\csname pct@rev.pol.rules.uiv.p2.generic-ppi-source@0.25.frozen.req\endcsname{188.6}
\expandafter\def\csname ppt@rev.pol.rules.uiv.p2.generic-ppi-source@0.25.frozen.req\endcsname{$+$188.6}
\expandafter\def\csname val@rev.pol.rules.uiv.p2.generic-ppi-source@0.25.frozen.score\endcsname{0.827}
\expandafter\def\csname cnt@rev.pol.rules.uiv.p2.generic-ppi-source@0.25.frozen.score\endcsname{1}
\expandafter\def\csname pct@rev.pol.rules.uiv.p2.generic-ppi-source@0.25.frozen.score\endcsname{82.7}
\expandafter\def\csname ppt@rev.pol.rules.uiv.p2.generic-ppi-source@0.25.frozen.score\endcsname{$+$82.7}
\expandafter\def\csname val@rev.pol.rules.uiv.p2.generic-ppi-source@0.25.lam0.5.keep\endcsname{0.544}
\expandafter\def\csname cnt@rev.pol.rules.uiv.p2.generic-ppi-source@0.25.lam0.5.keep\endcsname{1}
\expandafter\def\csname pct@rev.pol.rules.uiv.p2.generic-ppi-source@0.25.lam0.5.keep\endcsname{54.4}
\expandafter\def\csname ppt@rev.pol.rules.uiv.p2.generic-ppi-source@0.25.lam0.5.keep\endcsname{$+$54.4}
\expandafter\def\csname val@rev.pol.rules.uiv.p2.generic-ppi-source@0.25.lam0.5.req\endcsname{1.886}
\expandafter\def\csname cnt@rev.pol.rules.uiv.p2.generic-ppi-source@0.25.lam0.5.req\endcsname{2}
\expandafter\def\csname pct@rev.pol.rules.uiv.p2.generic-ppi-source@0.25.lam0.5.req\endcsname{188.6}
\expandafter\def\csname ppt@rev.pol.rules.uiv.p2.generic-ppi-source@0.25.lam0.5.req\endcsname{$+$188.6}
\expandafter\def\csname val@rev.pol.rules.uiv.p2.generic-ppi-source@0.25.lam0.5.score\endcsname{0.827}
\expandafter\def\csname cnt@rev.pol.rules.uiv.p2.generic-ppi-source@0.25.lam0.5.score\endcsname{1}
\expandafter\def\csname pct@rev.pol.rules.uiv.p2.generic-ppi-source@0.25.lam0.5.score\endcsname{82.7}
\expandafter\def\csname ppt@rev.pol.rules.uiv.p2.generic-ppi-source@0.25.lam0.5.score\endcsname{$+$82.7}
\expandafter\def\csname val@rev.pol.rules.uiv.p2.generic-ppi-source@0.25.lam1.keep\endcsname{0.413}
\expandafter\def\csname cnt@rev.pol.rules.uiv.p2.generic-ppi-source@0.25.lam1.keep\endcsname{0}
\expandafter\def\csname pct@rev.pol.rules.uiv.p2.generic-ppi-source@0.25.lam1.keep\endcsname{41.3}
\expandafter\def\csname ppt@rev.pol.rules.uiv.p2.generic-ppi-source@0.25.lam1.keep\endcsname{$+$41.3}
\expandafter\def\csname val@rev.pol.rules.uiv.p2.generic-ppi-source@0.25.lam1.req\endcsname{1.590}
\expandafter\def\csname cnt@rev.pol.rules.uiv.p2.generic-ppi-source@0.25.lam1.req\endcsname{2}
\expandafter\def\csname pct@rev.pol.rules.uiv.p2.generic-ppi-source@0.25.lam1.req\endcsname{159.0}
\expandafter\def\csname ppt@rev.pol.rules.uiv.p2.generic-ppi-source@0.25.lam1.req\endcsname{$+$159.0}
\expandafter\def\csname val@rev.pol.rules.uiv.p2.generic-ppi-source@0.25.lam1.score\endcsname{0.832}
\expandafter\def\csname cnt@rev.pol.rules.uiv.p2.generic-ppi-source@0.25.lam1.score\endcsname{1}
\expandafter\def\csname pct@rev.pol.rules.uiv.p2.generic-ppi-source@0.25.lam1.score\endcsname{83.2}
\expandafter\def\csname ppt@rev.pol.rules.uiv.p2.generic-ppi-source@0.25.lam1.score\endcsname{$+$83.2}
\expandafter\def\csname val@rev.pol.rules.uiv.p2.generic-ppi-source@0.25.m0.5.keep\endcsname{0.544}
\expandafter\def\csname cnt@rev.pol.rules.uiv.p2.generic-ppi-source@0.25.m0.5.keep\endcsname{1}
\expandafter\def\csname pct@rev.pol.rules.uiv.p2.generic-ppi-source@0.25.m0.5.keep\endcsname{54.4}
\expandafter\def\csname ppt@rev.pol.rules.uiv.p2.generic-ppi-source@0.25.m0.5.keep\endcsname{$+$54.4}
\expandafter\def\csname val@rev.pol.rules.uiv.p2.generic-ppi-source@0.25.m0.5.req\endcsname{1.886}
\expandafter\def\csname cnt@rev.pol.rules.uiv.p2.generic-ppi-source@0.25.m0.5.req\endcsname{2}
\expandafter\def\csname pct@rev.pol.rules.uiv.p2.generic-ppi-source@0.25.m0.5.req\endcsname{188.6}
\expandafter\def\csname ppt@rev.pol.rules.uiv.p2.generic-ppi-source@0.25.m0.5.req\endcsname{$+$188.6}
\expandafter\def\csname val@rev.pol.rules.uiv.p2.generic-ppi-source@0.25.m0.5.score\endcsname{0.827}
\expandafter\def\csname cnt@rev.pol.rules.uiv.p2.generic-ppi-source@0.25.m0.5.score\endcsname{1}
\expandafter\def\csname pct@rev.pol.rules.uiv.p2.generic-ppi-source@0.25.m0.5.score\endcsname{82.7}
\expandafter\def\csname ppt@rev.pol.rules.uiv.p2.generic-ppi-source@0.25.m0.5.score\endcsname{$+$82.7}
\expandafter\def\csname val@rev.pol.rules.uiv.p2.generic-ppi-source@0.25.m1.keep\endcsname{0.544}
\expandafter\def\csname cnt@rev.pol.rules.uiv.p2.generic-ppi-source@0.25.m1.keep\endcsname{1}
\expandafter\def\csname pct@rev.pol.rules.uiv.p2.generic-ppi-source@0.25.m1.keep\endcsname{54.4}
\expandafter\def\csname ppt@rev.pol.rules.uiv.p2.generic-ppi-source@0.25.m1.keep\endcsname{$+$54.4}
\expandafter\def\csname val@rev.pol.rules.uiv.p2.generic-ppi-source@0.25.m1.req\endcsname{1.886}
\expandafter\def\csname cnt@rev.pol.rules.uiv.p2.generic-ppi-source@0.25.m1.req\endcsname{2}
\expandafter\def\csname pct@rev.pol.rules.uiv.p2.generic-ppi-source@0.25.m1.req\endcsname{188.6}
\expandafter\def\csname ppt@rev.pol.rules.uiv.p2.generic-ppi-source@0.25.m1.req\endcsname{$+$188.6}
\expandafter\def\csname val@rev.pol.rules.uiv.p2.generic-ppi-source@0.25.m1.score\endcsname{0.827}
\expandafter\def\csname cnt@rev.pol.rules.uiv.p2.generic-ppi-source@0.25.m1.score\endcsname{1}
\expandafter\def\csname pct@rev.pol.rules.uiv.p2.generic-ppi-source@0.25.m1.score\endcsname{82.7}
\expandafter\def\csname ppt@rev.pol.rules.uiv.p2.generic-ppi-source@0.25.m1.score\endcsname{$+$82.7}
\expandafter\def\csname val@rev.pol.rules.uiv.p2.generic-ppi-source@0.25.m2.keep\endcsname{0.471}
\expandafter\def\csname cnt@rev.pol.rules.uiv.p2.generic-ppi-source@0.25.m2.keep\endcsname{0}
\expandafter\def\csname pct@rev.pol.rules.uiv.p2.generic-ppi-source@0.25.m2.keep\endcsname{47.1}
\expandafter\def\csname ppt@rev.pol.rules.uiv.p2.generic-ppi-source@0.25.m2.keep\endcsname{$+$47.1}
\expandafter\def\csname val@rev.pol.rules.uiv.p2.generic-ppi-source@0.25.m2.req\endcsname{1.739}
\expandafter\def\csname cnt@rev.pol.rules.uiv.p2.generic-ppi-source@0.25.m2.req\endcsname{2}
\expandafter\def\csname pct@rev.pol.rules.uiv.p2.generic-ppi-source@0.25.m2.req\endcsname{173.9}
\expandafter\def\csname ppt@rev.pol.rules.uiv.p2.generic-ppi-source@0.25.m2.req\endcsname{$+$173.9}
\expandafter\def\csname val@rev.pol.rules.uiv.p2.generic-ppi-source@0.25.m2.score\endcsname{0.832}
\expandafter\def\csname cnt@rev.pol.rules.uiv.p2.generic-ppi-source@0.25.m2.score\endcsname{1}
\expandafter\def\csname pct@rev.pol.rules.uiv.p2.generic-ppi-source@0.25.m2.score\endcsname{83.2}
\expandafter\def\csname ppt@rev.pol.rules.uiv.p2.generic-ppi-source@0.25.m2.score\endcsname{$+$83.2}
\expandafter\def\csname val@rev.pol.rules.uiv.p2.keep-all.keep\endcsname{1.000}
\expandafter\def\csname cnt@rev.pol.rules.uiv.p2.keep-all.keep\endcsname{1}
\expandafter\def\csname pct@rev.pol.rules.uiv.p2.keep-all.keep\endcsname{100.0}
\expandafter\def\csname ppt@rev.pol.rules.uiv.p2.keep-all.keep\endcsname{$+$100.0}
\expandafter\def\csname val@rev.pol.rules.uiv.p2.keep-all.req\endcsname{2.646}
\expandafter\def\csname cnt@rev.pol.rules.uiv.p2.keep-all.req\endcsname{3}
\expandafter\def\csname pct@rev.pol.rules.uiv.p2.keep-all.req\endcsname{264.6}
\expandafter\def\csname ppt@rev.pol.rules.uiv.p2.keep-all.req\endcsname{$+$264.6}
\expandafter\def\csname val@rev.pol.rules.uiv.p2.keep-all.score\endcsname{0.816}
\expandafter\def\csname cnt@rev.pol.rules.uiv.p2.keep-all.score\endcsname{1}
\expandafter\def\csname pct@rev.pol.rules.uiv.p2.keep-all.score\endcsname{81.6}
\expandafter\def\csname ppt@rev.pol.rules.uiv.p2.keep-all.score\endcsname{$+$81.6}
\expandafter\def\csname val@rev.pol.rules.uiv.p2.old-decision.frozen.keep\endcsname{0.500}
\expandafter\def\csname cnt@rev.pol.rules.uiv.p2.old-decision.frozen.keep\endcsname{0}
\expandafter\def\csname pct@rev.pol.rules.uiv.p2.old-decision.frozen.keep\endcsname{50.0}
\expandafter\def\csname ppt@rev.pol.rules.uiv.p2.old-decision.frozen.keep\endcsname{$+$50.0}
\expandafter\def\csname val@rev.pol.rules.uiv.p2.old-decision.frozen.req\endcsname{1.750}
\expandafter\def\csname cnt@rev.pol.rules.uiv.p2.old-decision.frozen.req\endcsname{2}
\expandafter\def\csname pct@rev.pol.rules.uiv.p2.old-decision.frozen.req\endcsname{175.0}
\expandafter\def\csname ppt@rev.pol.rules.uiv.p2.old-decision.frozen.req\endcsname{$+$175.0}
\expandafter\def\csname val@rev.pol.rules.uiv.p2.old-decision.frozen.score\endcsname{0.818}
\expandafter\def\csname cnt@rev.pol.rules.uiv.p2.old-decision.frozen.score\endcsname{1}
\expandafter\def\csname pct@rev.pol.rules.uiv.p2.old-decision.frozen.score\endcsname{81.8}
\expandafter\def\csname ppt@rev.pol.rules.uiv.p2.old-decision.frozen.score\endcsname{$+$81.8}
\expandafter\def\csname val@rev.pol.rules.uiv.p2.old-decision.lam0.5.keep\endcsname{0.250}
\expandafter\def\csname cnt@rev.pol.rules.uiv.p2.old-decision.lam0.5.keep\endcsname{0}
\expandafter\def\csname pct@rev.pol.rules.uiv.p2.old-decision.lam0.5.keep\endcsname{25.0}
\expandafter\def\csname ppt@rev.pol.rules.uiv.p2.old-decision.lam0.5.keep\endcsname{$+$25.0}
\expandafter\def\csname val@rev.pol.rules.uiv.p2.old-decision.lam0.5.req\endcsname{1.250}
\expandafter\def\csname cnt@rev.pol.rules.uiv.p2.old-decision.lam0.5.req\endcsname{1}
\expandafter\def\csname pct@rev.pol.rules.uiv.p2.old-decision.lam0.5.req\endcsname{125.0}
\expandafter\def\csname ppt@rev.pol.rules.uiv.p2.old-decision.lam0.5.req\endcsname{$+$125.0}
\expandafter\def\csname val@rev.pol.rules.uiv.p2.old-decision.lam0.5.score\endcsname{0.833}
\expandafter\def\csname cnt@rev.pol.rules.uiv.p2.old-decision.lam0.5.score\endcsname{1}
\expandafter\def\csname pct@rev.pol.rules.uiv.p2.old-decision.lam0.5.score\endcsname{83.3}
\expandafter\def\csname ppt@rev.pol.rules.uiv.p2.old-decision.lam0.5.score\endcsname{$+$83.3}
\expandafter\def\csname val@rev.pol.rules.uiv.p2.old-decision.lam1.keep\endcsname{0.250}
\expandafter\def\csname cnt@rev.pol.rules.uiv.p2.old-decision.lam1.keep\endcsname{0}
\expandafter\def\csname pct@rev.pol.rules.uiv.p2.old-decision.lam1.keep\endcsname{25.0}
\expandafter\def\csname ppt@rev.pol.rules.uiv.p2.old-decision.lam1.keep\endcsname{$+$25.0}
\expandafter\def\csname val@rev.pol.rules.uiv.p2.old-decision.lam1.req\endcsname{1.250}
\expandafter\def\csname cnt@rev.pol.rules.uiv.p2.old-decision.lam1.req\endcsname{1}
\expandafter\def\csname pct@rev.pol.rules.uiv.p2.old-decision.lam1.req\endcsname{125.0}
\expandafter\def\csname ppt@rev.pol.rules.uiv.p2.old-decision.lam1.req\endcsname{$+$125.0}
\expandafter\def\csname val@rev.pol.rules.uiv.p2.old-decision.lam1.score\endcsname{0.833}
\expandafter\def\csname cnt@rev.pol.rules.uiv.p2.old-decision.lam1.score\endcsname{1}
\expandafter\def\csname pct@rev.pol.rules.uiv.p2.old-decision.lam1.score\endcsname{83.3}
\expandafter\def\csname ppt@rev.pol.rules.uiv.p2.old-decision.lam1.score\endcsname{$+$83.3}
\expandafter\def\csname val@rev.pol.rules.uiv.p2.old-decision.m0.5.keep\endcsname{0.250}
\expandafter\def\csname cnt@rev.pol.rules.uiv.p2.old-decision.m0.5.keep\endcsname{0}
\expandafter\def\csname pct@rev.pol.rules.uiv.p2.old-decision.m0.5.keep\endcsname{25.0}
\expandafter\def\csname ppt@rev.pol.rules.uiv.p2.old-decision.m0.5.keep\endcsname{$+$25.0}
\expandafter\def\csname val@rev.pol.rules.uiv.p2.old-decision.m0.5.req\endcsname{1.250}
\expandafter\def\csname cnt@rev.pol.rules.uiv.p2.old-decision.m0.5.req\endcsname{1}
\expandafter\def\csname pct@rev.pol.rules.uiv.p2.old-decision.m0.5.req\endcsname{125.0}
\expandafter\def\csname ppt@rev.pol.rules.uiv.p2.old-decision.m0.5.req\endcsname{$+$125.0}
\expandafter\def\csname val@rev.pol.rules.uiv.p2.old-decision.m0.5.score\endcsname{0.833}
\expandafter\def\csname cnt@rev.pol.rules.uiv.p2.old-decision.m0.5.score\endcsname{1}
\expandafter\def\csname pct@rev.pol.rules.uiv.p2.old-decision.m0.5.score\endcsname{83.3}
\expandafter\def\csname ppt@rev.pol.rules.uiv.p2.old-decision.m0.5.score\endcsname{$+$83.3}
\expandafter\def\csname val@rev.pol.rules.uiv.p2.old-decision.m1.keep\endcsname{0.250}
\expandafter\def\csname cnt@rev.pol.rules.uiv.p2.old-decision.m1.keep\endcsname{0}
\expandafter\def\csname pct@rev.pol.rules.uiv.p2.old-decision.m1.keep\endcsname{25.0}
\expandafter\def\csname ppt@rev.pol.rules.uiv.p2.old-decision.m1.keep\endcsname{$+$25.0}
\expandafter\def\csname val@rev.pol.rules.uiv.p2.old-decision.m1.req\endcsname{1.250}
\expandafter\def\csname cnt@rev.pol.rules.uiv.p2.old-decision.m1.req\endcsname{1}
\expandafter\def\csname pct@rev.pol.rules.uiv.p2.old-decision.m1.req\endcsname{125.0}
\expandafter\def\csname ppt@rev.pol.rules.uiv.p2.old-decision.m1.req\endcsname{$+$125.0}
\expandafter\def\csname val@rev.pol.rules.uiv.p2.old-decision.m1.score\endcsname{0.833}
\expandafter\def\csname cnt@rev.pol.rules.uiv.p2.old-decision.m1.score\endcsname{1}
\expandafter\def\csname pct@rev.pol.rules.uiv.p2.old-decision.m1.score\endcsname{83.3}
\expandafter\def\csname ppt@rev.pol.rules.uiv.p2.old-decision.m1.score\endcsname{$+$83.3}
\expandafter\def\csname val@rev.pol.rules.uiv.p2.old-decision.m2.keep\endcsname{0.000}
\expandafter\def\csname cnt@rev.pol.rules.uiv.p2.old-decision.m2.keep\endcsname{0}
\expandafter\def\csname pct@rev.pol.rules.uiv.p2.old-decision.m2.keep\endcsname{0.0}
\expandafter\def\csname ppt@rev.pol.rules.uiv.p2.old-decision.m2.keep\endcsname{0.0}
\expandafter\def\csname val@rev.pol.rules.uiv.p2.old-decision.m2.req\endcsname{1.000}
\expandafter\def\csname cnt@rev.pol.rules.uiv.p2.old-decision.m2.req\endcsname{1}
\expandafter\def\csname pct@rev.pol.rules.uiv.p2.old-decision.m2.req\endcsname{100.0}
\expandafter\def\csname ppt@rev.pol.rules.uiv.p2.old-decision.m2.req\endcsname{$+$100.0}
\expandafter\def\csname val@rev.pol.rules.uiv.p2.old-decision.m2.score\endcsname{0.840}
\expandafter\def\csname cnt@rev.pol.rules.uiv.p2.old-decision.m2.score\endcsname{1}
\expandafter\def\csname pct@rev.pol.rules.uiv.p2.old-decision.m2.score\endcsname{84.0}
\expandafter\def\csname ppt@rev.pol.rules.uiv.p2.old-decision.m2.score\endcsname{$+$84.0}
\expandafter\def\csname val@rev.pol.rules.uiv.p2.oracle-on-T.keep\endcsname{0.250}
\expandafter\def\csname cnt@rev.pol.rules.uiv.p2.oracle-on-T.keep\endcsname{0}
\expandafter\def\csname pct@rev.pol.rules.uiv.p2.oracle-on-T.keep\endcsname{25.0}
\expandafter\def\csname ppt@rev.pol.rules.uiv.p2.oracle-on-T.keep\endcsname{$+$25.0}
\expandafter\def\csname val@rev.pol.rules.uiv.p2.oracle-on-T.req\endcsname{1.250}
\expandafter\def\csname cnt@rev.pol.rules.uiv.p2.oracle-on-T.req\endcsname{1}
\expandafter\def\csname pct@rev.pol.rules.uiv.p2.oracle-on-T.req\endcsname{125.0}
\expandafter\def\csname ppt@rev.pol.rules.uiv.p2.oracle-on-T.req\endcsname{$+$125.0}
\expandafter\def\csname val@rev.pol.rules.uiv.p2.oracle-on-T.score\endcsname{0.840}
\expandafter\def\csname cnt@rev.pol.rules.uiv.p2.oracle-on-T.score\endcsname{1}
\expandafter\def\csname pct@rev.pol.rules.uiv.p2.oracle-on-T.score\endcsname{84.0}
\expandafter\def\csname ppt@rev.pol.rules.uiv.p2.oracle-on-T.score\endcsname{$+$84.0}
\expandafter\def\csname val@rev.pol.rules.uiv.p2.r-s.center-d1\endcsname{2.000}
\expandafter\def\csname cnt@rev.pol.rules.uiv.p2.r-s.center-d1\endcsname{2}
\expandafter\def\csname pct@rev.pol.rules.uiv.p2.r-s.center-d1\endcsname{200.0}
\expandafter\def\csname ppt@rev.pol.rules.uiv.p2.r-s.center-d1\endcsname{$+$200.0}
\expandafter\def\csname val@rev.pol.rules.uiv.p2.r-s.context-d1\endcsname{2.000}
\expandafter\def\csname cnt@rev.pol.rules.uiv.p2.r-s.context-d1\endcsname{2}
\expandafter\def\csname pct@rev.pol.rules.uiv.p2.r-s.context-d1\endcsname{200.0}
\expandafter\def\csname ppt@rev.pol.rules.uiv.p2.r-s.context-d1\endcsname{$+$200.0}
\expandafter\def\csname val@rev.pol.rules.uiv.p2.r-s.dual-d2\endcsname{3.000}
\expandafter\def\csname cnt@rev.pol.rules.uiv.p2.r-s.dual-d2\endcsname{3}
\expandafter\def\csname pct@rev.pol.rules.uiv.p2.r-s.dual-d2\endcsname{300.0}
\expandafter\def\csname ppt@rev.pol.rules.uiv.p2.r-s.dual-d2\endcsname{$+$300.0}
\expandafter\def\csname val@rev.pol.rules.uiv.p2.r-s.grid2-d3\endcsname{3.583}
\expandafter\def\csname cnt@rev.pol.rules.uiv.p2.r-s.grid2-d3\endcsname{4}
\expandafter\def\csname pct@rev.pol.rules.uiv.p2.r-s.grid2-d3\endcsname{358.3}
\expandafter\def\csname ppt@rev.pol.rules.uiv.p2.r-s.grid2-d3\endcsname{$+$358.3}
\expandafter\def\csname val@rev.pol.rules.uiv.p2.rcr-stratified@0.25.frozen.keep\endcsname{0.588}
\expandafter\def\csname cnt@rev.pol.rules.uiv.p2.rcr-stratified@0.25.frozen.keep\endcsname{1}
\expandafter\def\csname pct@rev.pol.rules.uiv.p2.rcr-stratified@0.25.frozen.keep\endcsname{58.8}
\expandafter\def\csname ppt@rev.pol.rules.uiv.p2.rcr-stratified@0.25.frozen.keep\endcsname{$+$58.8}
\expandafter\def\csname val@rev.pol.rules.uiv.p2.rcr-stratified@0.25.frozen.req\endcsname{1.911}
\expandafter\def\csname cnt@rev.pol.rules.uiv.p2.rcr-stratified@0.25.frozen.req\endcsname{2}
\expandafter\def\csname pct@rev.pol.rules.uiv.p2.rcr-stratified@0.25.frozen.req\endcsname{191.1}
\expandafter\def\csname ppt@rev.pol.rules.uiv.p2.rcr-stratified@0.25.frozen.req\endcsname{$+$191.1}
\expandafter\def\csname val@rev.pol.rules.uiv.p2.rcr-stratified@0.25.frozen.score\endcsname{0.828}
\expandafter\def\csname cnt@rev.pol.rules.uiv.p2.rcr-stratified@0.25.frozen.score\endcsname{1}
\expandafter\def\csname pct@rev.pol.rules.uiv.p2.rcr-stratified@0.25.frozen.score\endcsname{82.8}
\expandafter\def\csname ppt@rev.pol.rules.uiv.p2.rcr-stratified@0.25.frozen.score\endcsname{$+$82.8}
\expandafter\def\csname val@rev.pol.rules.uiv.p2.rcr-stratified@0.25.lam0.5.keep\endcsname{0.530}
\expandafter\def\csname cnt@rev.pol.rules.uiv.p2.rcr-stratified@0.25.lam0.5.keep\endcsname{1}
\expandafter\def\csname pct@rev.pol.rules.uiv.p2.rcr-stratified@0.25.lam0.5.keep\endcsname{53.0}
\expandafter\def\csname ppt@rev.pol.rules.uiv.p2.rcr-stratified@0.25.lam0.5.keep\endcsname{$+$53.0}
\expandafter\def\csname val@rev.pol.rules.uiv.p2.rcr-stratified@0.25.lam0.5.req\endcsname{1.772}
\expandafter\def\csname cnt@rev.pol.rules.uiv.p2.rcr-stratified@0.25.lam0.5.req\endcsname{2}
\expandafter\def\csname pct@rev.pol.rules.uiv.p2.rcr-stratified@0.25.lam0.5.req\endcsname{177.2}
\expandafter\def\csname ppt@rev.pol.rules.uiv.p2.rcr-stratified@0.25.lam0.5.req\endcsname{$+$177.2}
\expandafter\def\csname val@rev.pol.rules.uiv.p2.rcr-stratified@0.25.lam0.5.score\endcsname{0.830}
\expandafter\def\csname cnt@rev.pol.rules.uiv.p2.rcr-stratified@0.25.lam0.5.score\endcsname{1}
\expandafter\def\csname pct@rev.pol.rules.uiv.p2.rcr-stratified@0.25.lam0.5.score\endcsname{83.0}
\expandafter\def\csname ppt@rev.pol.rules.uiv.p2.rcr-stratified@0.25.lam0.5.score\endcsname{$+$83.0}
\expandafter\def\csname val@rev.pol.rules.uiv.p2.rcr-stratified@0.25.lam1.keep\endcsname{0.446}
\expandafter\def\csname cnt@rev.pol.rules.uiv.p2.rcr-stratified@0.25.lam1.keep\endcsname{0}
\expandafter\def\csname pct@rev.pol.rules.uiv.p2.rcr-stratified@0.25.lam1.keep\endcsname{44.6}
\expandafter\def\csname ppt@rev.pol.rules.uiv.p2.rcr-stratified@0.25.lam1.keep\endcsname{$+$44.6}
\expandafter\def\csname val@rev.pol.rules.uiv.p2.rcr-stratified@0.25.lam1.req\endcsname{1.656}
\expandafter\def\csname cnt@rev.pol.rules.uiv.p2.rcr-stratified@0.25.lam1.req\endcsname{2}
\expandafter\def\csname pct@rev.pol.rules.uiv.p2.rcr-stratified@0.25.lam1.req\endcsname{165.6}
\expandafter\def\csname ppt@rev.pol.rules.uiv.p2.rcr-stratified@0.25.lam1.req\endcsname{$+$165.6}
\expandafter\def\csname val@rev.pol.rules.uiv.p2.rcr-stratified@0.25.lam1.score\endcsname{0.830}
\expandafter\def\csname cnt@rev.pol.rules.uiv.p2.rcr-stratified@0.25.lam1.score\endcsname{1}
\expandafter\def\csname pct@rev.pol.rules.uiv.p2.rcr-stratified@0.25.lam1.score\endcsname{83.0}
\expandafter\def\csname ppt@rev.pol.rules.uiv.p2.rcr-stratified@0.25.lam1.score\endcsname{$+$83.0}
\expandafter\def\csname val@rev.pol.rules.uiv.p2.rcr-stratified@0.25.m0.5.keep\endcsname{0.530}
\expandafter\def\csname cnt@rev.pol.rules.uiv.p2.rcr-stratified@0.25.m0.5.keep\endcsname{1}
\expandafter\def\csname pct@rev.pol.rules.uiv.p2.rcr-stratified@0.25.m0.5.keep\endcsname{53.0}
\expandafter\def\csname ppt@rev.pol.rules.uiv.p2.rcr-stratified@0.25.m0.5.keep\endcsname{$+$53.0}
\expandafter\def\csname val@rev.pol.rules.uiv.p2.rcr-stratified@0.25.m0.5.req\endcsname{1.772}
\expandafter\def\csname cnt@rev.pol.rules.uiv.p2.rcr-stratified@0.25.m0.5.req\endcsname{2}
\expandafter\def\csname pct@rev.pol.rules.uiv.p2.rcr-stratified@0.25.m0.5.req\endcsname{177.2}
\expandafter\def\csname ppt@rev.pol.rules.uiv.p2.rcr-stratified@0.25.m0.5.req\endcsname{$+$177.2}
\expandafter\def\csname val@rev.pol.rules.uiv.p2.rcr-stratified@0.25.m0.5.score\endcsname{0.830}
\expandafter\def\csname cnt@rev.pol.rules.uiv.p2.rcr-stratified@0.25.m0.5.score\endcsname{1}
\expandafter\def\csname pct@rev.pol.rules.uiv.p2.rcr-stratified@0.25.m0.5.score\endcsname{83.0}
\expandafter\def\csname ppt@rev.pol.rules.uiv.p2.rcr-stratified@0.25.m0.5.score\endcsname{$+$83.0}
\expandafter\def\csname val@rev.pol.rules.uiv.p2.rcr-stratified@0.25.m1.keep\endcsname{0.466}
\expandafter\def\csname cnt@rev.pol.rules.uiv.p2.rcr-stratified@0.25.m1.keep\endcsname{0}
\expandafter\def\csname pct@rev.pol.rules.uiv.p2.rcr-stratified@0.25.m1.keep\endcsname{46.6}
\expandafter\def\csname ppt@rev.pol.rules.uiv.p2.rcr-stratified@0.25.m1.keep\endcsname{$+$46.6}
\expandafter\def\csname val@rev.pol.rules.uiv.p2.rcr-stratified@0.25.m1.req\endcsname{1.707}
\expandafter\def\csname cnt@rev.pol.rules.uiv.p2.rcr-stratified@0.25.m1.req\endcsname{2}
\expandafter\def\csname pct@rev.pol.rules.uiv.p2.rcr-stratified@0.25.m1.req\endcsname{170.7}
\expandafter\def\csname ppt@rev.pol.rules.uiv.p2.rcr-stratified@0.25.m1.req\endcsname{$+$170.7}
\expandafter\def\csname val@rev.pol.rules.uiv.p2.rcr-stratified@0.25.m1.score\endcsname{0.830}
\expandafter\def\csname cnt@rev.pol.rules.uiv.p2.rcr-stratified@0.25.m1.score\endcsname{1}
\expandafter\def\csname pct@rev.pol.rules.uiv.p2.rcr-stratified@0.25.m1.score\endcsname{83.0}
\expandafter\def\csname ppt@rev.pol.rules.uiv.p2.rcr-stratified@0.25.m1.score\endcsname{$+$83.0}
\expandafter\def\csname val@rev.pol.rules.uiv.p2.rcr-stratified@0.25.m2.keep\endcsname{0.400}
\expandafter\def\csname cnt@rev.pol.rules.uiv.p2.rcr-stratified@0.25.m2.keep\endcsname{0}
\expandafter\def\csname pct@rev.pol.rules.uiv.p2.rcr-stratified@0.25.m2.keep\endcsname{40.0}
\expandafter\def\csname ppt@rev.pol.rules.uiv.p2.rcr-stratified@0.25.m2.keep\endcsname{$+$40.0}
\expandafter\def\csname val@rev.pol.rules.uiv.p2.rcr-stratified@0.25.m2.req\endcsname{1.610}
\expandafter\def\csname cnt@rev.pol.rules.uiv.p2.rcr-stratified@0.25.m2.req\endcsname{2}
\expandafter\def\csname pct@rev.pol.rules.uiv.p2.rcr-stratified@0.25.m2.req\endcsname{161.0}
\expandafter\def\csname ppt@rev.pol.rules.uiv.p2.rcr-stratified@0.25.m2.req\endcsname{$+$161.0}
\expandafter\def\csname val@rev.pol.rules.uiv.p2.rcr-stratified@0.25.m2.score\endcsname{0.831}
\expandafter\def\csname cnt@rev.pol.rules.uiv.p2.rcr-stratified@0.25.m2.score\endcsname{1}
\expandafter\def\csname pct@rev.pol.rules.uiv.p2.rcr-stratified@0.25.m2.score\endcsname{83.1}
\expandafter\def\csname ppt@rev.pol.rules.uiv.p2.rcr-stratified@0.25.m2.score\endcsname{$+$83.1}
\expandafter\def\csname val@rev.pol.rules.uiv.p2.replay-only.frozen.keep\endcsname{0.500}
\expandafter\def\csname cnt@rev.pol.rules.uiv.p2.replay-only.frozen.keep\endcsname{0}
\expandafter\def\csname pct@rev.pol.rules.uiv.p2.replay-only.frozen.keep\endcsname{50.0}
\expandafter\def\csname ppt@rev.pol.rules.uiv.p2.replay-only.frozen.keep\endcsname{$+$50.0}
\expandafter\def\csname val@rev.pol.rules.uiv.p2.replay-only.frozen.req\endcsname{1.500}
\expandafter\def\csname cnt@rev.pol.rules.uiv.p2.replay-only.frozen.req\endcsname{2}
\expandafter\def\csname pct@rev.pol.rules.uiv.p2.replay-only.frozen.req\endcsname{150.0}
\expandafter\def\csname ppt@rev.pol.rules.uiv.p2.replay-only.frozen.req\endcsname{$+$150.0}
\expandafter\def\csname val@rev.pol.rules.uiv.p2.replay-only.frozen.score\endcsname{0.833}
\expandafter\def\csname cnt@rev.pol.rules.uiv.p2.replay-only.frozen.score\endcsname{1}
\expandafter\def\csname pct@rev.pol.rules.uiv.p2.replay-only.frozen.score\endcsname{83.3}
\expandafter\def\csname ppt@rev.pol.rules.uiv.p2.replay-only.frozen.score\endcsname{$+$83.3}
\expandafter\def\csname val@rev.pol.rules.uiv.p2.replay-only.lam0.5.keep\endcsname{0.500}
\expandafter\def\csname cnt@rev.pol.rules.uiv.p2.replay-only.lam0.5.keep\endcsname{0}
\expandafter\def\csname pct@rev.pol.rules.uiv.p2.replay-only.lam0.5.keep\endcsname{50.0}
\expandafter\def\csname ppt@rev.pol.rules.uiv.p2.replay-only.lam0.5.keep\endcsname{$+$50.0}
\expandafter\def\csname val@rev.pol.rules.uiv.p2.replay-only.lam0.5.req\endcsname{1.500}
\expandafter\def\csname cnt@rev.pol.rules.uiv.p2.replay-only.lam0.5.req\endcsname{2}
\expandafter\def\csname pct@rev.pol.rules.uiv.p2.replay-only.lam0.5.req\endcsname{150.0}
\expandafter\def\csname ppt@rev.pol.rules.uiv.p2.replay-only.lam0.5.req\endcsname{$+$150.0}
\expandafter\def\csname val@rev.pol.rules.uiv.p2.replay-only.lam0.5.score\endcsname{0.833}
\expandafter\def\csname cnt@rev.pol.rules.uiv.p2.replay-only.lam0.5.score\endcsname{1}
\expandafter\def\csname pct@rev.pol.rules.uiv.p2.replay-only.lam0.5.score\endcsname{83.3}
\expandafter\def\csname ppt@rev.pol.rules.uiv.p2.replay-only.lam0.5.score\endcsname{$+$83.3}
\expandafter\def\csname val@rev.pol.rules.uiv.p2.replay-only.lam1.keep\endcsname{0.500}
\expandafter\def\csname cnt@rev.pol.rules.uiv.p2.replay-only.lam1.keep\endcsname{0}
\expandafter\def\csname pct@rev.pol.rules.uiv.p2.replay-only.lam1.keep\endcsname{50.0}
\expandafter\def\csname ppt@rev.pol.rules.uiv.p2.replay-only.lam1.keep\endcsname{$+$50.0}
\expandafter\def\csname val@rev.pol.rules.uiv.p2.replay-only.lam1.req\endcsname{1.500}
\expandafter\def\csname cnt@rev.pol.rules.uiv.p2.replay-only.lam1.req\endcsname{2}
\expandafter\def\csname pct@rev.pol.rules.uiv.p2.replay-only.lam1.req\endcsname{150.0}
\expandafter\def\csname ppt@rev.pol.rules.uiv.p2.replay-only.lam1.req\endcsname{$+$150.0}
\expandafter\def\csname val@rev.pol.rules.uiv.p2.replay-only.lam1.score\endcsname{0.833}
\expandafter\def\csname cnt@rev.pol.rules.uiv.p2.replay-only.lam1.score\endcsname{1}
\expandafter\def\csname pct@rev.pol.rules.uiv.p2.replay-only.lam1.score\endcsname{83.3}
\expandafter\def\csname ppt@rev.pol.rules.uiv.p2.replay-only.lam1.score\endcsname{$+$83.3}
\expandafter\def\csname val@rev.pol.rules.uiv.p2.replay-only.m0.5.keep\endcsname{0.500}
\expandafter\def\csname cnt@rev.pol.rules.uiv.p2.replay-only.m0.5.keep\endcsname{0}
\expandafter\def\csname pct@rev.pol.rules.uiv.p2.replay-only.m0.5.keep\endcsname{50.0}
\expandafter\def\csname ppt@rev.pol.rules.uiv.p2.replay-only.m0.5.keep\endcsname{$+$50.0}
\expandafter\def\csname val@rev.pol.rules.uiv.p2.replay-only.m0.5.req\endcsname{1.500}
\expandafter\def\csname cnt@rev.pol.rules.uiv.p2.replay-only.m0.5.req\endcsname{2}
\expandafter\def\csname pct@rev.pol.rules.uiv.p2.replay-only.m0.5.req\endcsname{150.0}
\expandafter\def\csname ppt@rev.pol.rules.uiv.p2.replay-only.m0.5.req\endcsname{$+$150.0}
\expandafter\def\csname val@rev.pol.rules.uiv.p2.replay-only.m0.5.score\endcsname{0.833}
\expandafter\def\csname cnt@rev.pol.rules.uiv.p2.replay-only.m0.5.score\endcsname{1}
\expandafter\def\csname pct@rev.pol.rules.uiv.p2.replay-only.m0.5.score\endcsname{83.3}
\expandafter\def\csname ppt@rev.pol.rules.uiv.p2.replay-only.m0.5.score\endcsname{$+$83.3}
\expandafter\def\csname val@rev.pol.rules.uiv.p2.replay-only.m1.keep\endcsname{0.500}
\expandafter\def\csname cnt@rev.pol.rules.uiv.p2.replay-only.m1.keep\endcsname{0}
\expandafter\def\csname pct@rev.pol.rules.uiv.p2.replay-only.m1.keep\endcsname{50.0}
\expandafter\def\csname ppt@rev.pol.rules.uiv.p2.replay-only.m1.keep\endcsname{$+$50.0}
\expandafter\def\csname val@rev.pol.rules.uiv.p2.replay-only.m1.req\endcsname{1.500}
\expandafter\def\csname cnt@rev.pol.rules.uiv.p2.replay-only.m1.req\endcsname{2}
\expandafter\def\csname pct@rev.pol.rules.uiv.p2.replay-only.m1.req\endcsname{150.0}
\expandafter\def\csname ppt@rev.pol.rules.uiv.p2.replay-only.m1.req\endcsname{$+$150.0}
\expandafter\def\csname val@rev.pol.rules.uiv.p2.replay-only.m1.score\endcsname{0.833}
\expandafter\def\csname cnt@rev.pol.rules.uiv.p2.replay-only.m1.score\endcsname{1}
\expandafter\def\csname pct@rev.pol.rules.uiv.p2.replay-only.m1.score\endcsname{83.3}
\expandafter\def\csname ppt@rev.pol.rules.uiv.p2.replay-only.m1.score\endcsname{$+$83.3}
\expandafter\def\csname val@rev.pol.rules.uiv.p2.replay-only.m2.keep\endcsname{0.250}
\expandafter\def\csname cnt@rev.pol.rules.uiv.p2.replay-only.m2.keep\endcsname{0}
\expandafter\def\csname pct@rev.pol.rules.uiv.p2.replay-only.m2.keep\endcsname{25.0}
\expandafter\def\csname ppt@rev.pol.rules.uiv.p2.replay-only.m2.keep\endcsname{$+$25.0}
\expandafter\def\csname val@rev.pol.rules.uiv.p2.replay-only.m2.req\endcsname{1.250}
\expandafter\def\csname cnt@rev.pol.rules.uiv.p2.replay-only.m2.req\endcsname{1}
\expandafter\def\csname pct@rev.pol.rules.uiv.p2.replay-only.m2.req\endcsname{125.0}
\expandafter\def\csname ppt@rev.pol.rules.uiv.p2.replay-only.m2.req\endcsname{$+$125.0}
\expandafter\def\csname val@rev.pol.rules.uiv.p2.replay-only.m2.score\endcsname{0.840}
\expandafter\def\csname cnt@rev.pol.rules.uiv.p2.replay-only.m2.score\endcsname{1}
\expandafter\def\csname pct@rev.pol.rules.uiv.p2.replay-only.m2.score\endcsname{84.0}
\expandafter\def\csname ppt@rev.pol.rules.uiv.p2.replay-only.m2.score\endcsname{$+$84.0}
\expandafter\def\csname val@rev.pol.rules.uiv.p2.replay-ppi-uniform@0.25.frozen.keep\endcsname{0.580}
\expandafter\def\csname cnt@rev.pol.rules.uiv.p2.replay-ppi-uniform@0.25.frozen.keep\endcsname{1}
\expandafter\def\csname pct@rev.pol.rules.uiv.p2.replay-ppi-uniform@0.25.frozen.keep\endcsname{58.0}
\expandafter\def\csname ppt@rev.pol.rules.uiv.p2.replay-ppi-uniform@0.25.frozen.keep\endcsname{$+$58.0}
\expandafter\def\csname val@rev.pol.rules.uiv.p2.replay-ppi-uniform@0.25.frozen.req\endcsname{1.908}
\expandafter\def\csname cnt@rev.pol.rules.uiv.p2.replay-ppi-uniform@0.25.frozen.req\endcsname{2}
\expandafter\def\csname pct@rev.pol.rules.uiv.p2.replay-ppi-uniform@0.25.frozen.req\endcsname{190.8}
\expandafter\def\csname ppt@rev.pol.rules.uiv.p2.replay-ppi-uniform@0.25.frozen.req\endcsname{$+$190.8}
\expandafter\def\csname val@rev.pol.rules.uiv.p2.replay-ppi-uniform@0.25.frozen.score\endcsname{0.827}
\expandafter\def\csname cnt@rev.pol.rules.uiv.p2.replay-ppi-uniform@0.25.frozen.score\endcsname{1}
\expandafter\def\csname pct@rev.pol.rules.uiv.p2.replay-ppi-uniform@0.25.frozen.score\endcsname{82.7}
\expandafter\def\csname ppt@rev.pol.rules.uiv.p2.replay-ppi-uniform@0.25.frozen.score\endcsname{$+$82.7}
\expandafter\def\csname val@rev.pol.rules.uiv.p2.replay-ppi-uniform@0.25.lam0.5.keep\endcsname{0.580}
\expandafter\def\csname cnt@rev.pol.rules.uiv.p2.replay-ppi-uniform@0.25.lam0.5.keep\endcsname{1}
\expandafter\def\csname pct@rev.pol.rules.uiv.p2.replay-ppi-uniform@0.25.lam0.5.keep\endcsname{58.0}
\expandafter\def\csname ppt@rev.pol.rules.uiv.p2.replay-ppi-uniform@0.25.lam0.5.keep\endcsname{$+$58.0}
\expandafter\def\csname val@rev.pol.rules.uiv.p2.replay-ppi-uniform@0.25.lam0.5.req\endcsname{1.908}
\expandafter\def\csname cnt@rev.pol.rules.uiv.p2.replay-ppi-uniform@0.25.lam0.5.req\endcsname{2}
\expandafter\def\csname pct@rev.pol.rules.uiv.p2.replay-ppi-uniform@0.25.lam0.5.req\endcsname{190.8}
\expandafter\def\csname ppt@rev.pol.rules.uiv.p2.replay-ppi-uniform@0.25.lam0.5.req\endcsname{$+$190.8}
\expandafter\def\csname val@rev.pol.rules.uiv.p2.replay-ppi-uniform@0.25.lam0.5.score\endcsname{0.827}
\expandafter\def\csname cnt@rev.pol.rules.uiv.p2.replay-ppi-uniform@0.25.lam0.5.score\endcsname{1}
\expandafter\def\csname pct@rev.pol.rules.uiv.p2.replay-ppi-uniform@0.25.lam0.5.score\endcsname{82.7}
\expandafter\def\csname ppt@rev.pol.rules.uiv.p2.replay-ppi-uniform@0.25.lam0.5.score\endcsname{$+$82.7}
\expandafter\def\csname val@rev.pol.rules.uiv.p2.replay-ppi-uniform@0.25.lam1.keep\endcsname{0.517}
\expandafter\def\csname cnt@rev.pol.rules.uiv.p2.replay-ppi-uniform@0.25.lam1.keep\endcsname{1}
\expandafter\def\csname pct@rev.pol.rules.uiv.p2.replay-ppi-uniform@0.25.lam1.keep\endcsname{51.7}
\expandafter\def\csname ppt@rev.pol.rules.uiv.p2.replay-ppi-uniform@0.25.lam1.keep\endcsname{$+$51.7}
\expandafter\def\csname val@rev.pol.rules.uiv.p2.replay-ppi-uniform@0.25.lam1.req\endcsname{1.748}
\expandafter\def\csname cnt@rev.pol.rules.uiv.p2.replay-ppi-uniform@0.25.lam1.req\endcsname{2}
\expandafter\def\csname pct@rev.pol.rules.uiv.p2.replay-ppi-uniform@0.25.lam1.req\endcsname{174.8}
\expandafter\def\csname ppt@rev.pol.rules.uiv.p2.replay-ppi-uniform@0.25.lam1.req\endcsname{$+$174.8}
\expandafter\def\csname val@rev.pol.rules.uiv.p2.replay-ppi-uniform@0.25.lam1.score\endcsname{0.828}
\expandafter\def\csname cnt@rev.pol.rules.uiv.p2.replay-ppi-uniform@0.25.lam1.score\endcsname{1}
\expandafter\def\csname pct@rev.pol.rules.uiv.p2.replay-ppi-uniform@0.25.lam1.score\endcsname{82.8}
\expandafter\def\csname ppt@rev.pol.rules.uiv.p2.replay-ppi-uniform@0.25.lam1.score\endcsname{$+$82.8}
\expandafter\def\csname val@rev.pol.rules.uiv.p2.replay-ppi-uniform@0.25.m0.5.keep\endcsname{0.580}
\expandafter\def\csname cnt@rev.pol.rules.uiv.p2.replay-ppi-uniform@0.25.m0.5.keep\endcsname{1}
\expandafter\def\csname pct@rev.pol.rules.uiv.p2.replay-ppi-uniform@0.25.m0.5.keep\endcsname{58.0}
\expandafter\def\csname ppt@rev.pol.rules.uiv.p2.replay-ppi-uniform@0.25.m0.5.keep\endcsname{$+$58.0}
\expandafter\def\csname val@rev.pol.rules.uiv.p2.replay-ppi-uniform@0.25.m0.5.req\endcsname{1.908}
\expandafter\def\csname cnt@rev.pol.rules.uiv.p2.replay-ppi-uniform@0.25.m0.5.req\endcsname{2}
\expandafter\def\csname pct@rev.pol.rules.uiv.p2.replay-ppi-uniform@0.25.m0.5.req\endcsname{190.8}
\expandafter\def\csname ppt@rev.pol.rules.uiv.p2.replay-ppi-uniform@0.25.m0.5.req\endcsname{$+$190.8}
\expandafter\def\csname val@rev.pol.rules.uiv.p2.replay-ppi-uniform@0.25.m0.5.score\endcsname{0.827}
\expandafter\def\csname cnt@rev.pol.rules.uiv.p2.replay-ppi-uniform@0.25.m0.5.score\endcsname{1}
\expandafter\def\csname pct@rev.pol.rules.uiv.p2.replay-ppi-uniform@0.25.m0.5.score\endcsname{82.7}
\expandafter\def\csname ppt@rev.pol.rules.uiv.p2.replay-ppi-uniform@0.25.m0.5.score\endcsname{$+$82.7}
\expandafter\def\csname val@rev.pol.rules.uiv.p2.replay-ppi-uniform@0.25.m1.keep\endcsname{0.580}
\expandafter\def\csname cnt@rev.pol.rules.uiv.p2.replay-ppi-uniform@0.25.m1.keep\endcsname{1}
\expandafter\def\csname pct@rev.pol.rules.uiv.p2.replay-ppi-uniform@0.25.m1.keep\endcsname{58.0}
\expandafter\def\csname ppt@rev.pol.rules.uiv.p2.replay-ppi-uniform@0.25.m1.keep\endcsname{$+$58.0}
\expandafter\def\csname val@rev.pol.rules.uiv.p2.replay-ppi-uniform@0.25.m1.req\endcsname{1.908}
\expandafter\def\csname cnt@rev.pol.rules.uiv.p2.replay-ppi-uniform@0.25.m1.req\endcsname{2}
\expandafter\def\csname pct@rev.pol.rules.uiv.p2.replay-ppi-uniform@0.25.m1.req\endcsname{190.8}
\expandafter\def\csname ppt@rev.pol.rules.uiv.p2.replay-ppi-uniform@0.25.m1.req\endcsname{$+$190.8}
\expandafter\def\csname val@rev.pol.rules.uiv.p2.replay-ppi-uniform@0.25.m1.score\endcsname{0.827}
\expandafter\def\csname cnt@rev.pol.rules.uiv.p2.replay-ppi-uniform@0.25.m1.score\endcsname{1}
\expandafter\def\csname pct@rev.pol.rules.uiv.p2.replay-ppi-uniform@0.25.m1.score\endcsname{82.7}
\expandafter\def\csname ppt@rev.pol.rules.uiv.p2.replay-ppi-uniform@0.25.m1.score\endcsname{$+$82.7}
\expandafter\def\csname val@rev.pol.rules.uiv.p2.replay-ppi-uniform@0.25.m2.keep\endcsname{0.493}
\expandafter\def\csname cnt@rev.pol.rules.uiv.p2.replay-ppi-uniform@0.25.m2.keep\endcsname{0}
\expandafter\def\csname pct@rev.pol.rules.uiv.p2.replay-ppi-uniform@0.25.m2.keep\endcsname{49.3}
\expandafter\def\csname ppt@rev.pol.rules.uiv.p2.replay-ppi-uniform@0.25.m2.keep\endcsname{$+$49.3}
\expandafter\def\csname val@rev.pol.rules.uiv.p2.replay-ppi-uniform@0.25.m2.req\endcsname{1.822}
\expandafter\def\csname cnt@rev.pol.rules.uiv.p2.replay-ppi-uniform@0.25.m2.req\endcsname{2}
\expandafter\def\csname pct@rev.pol.rules.uiv.p2.replay-ppi-uniform@0.25.m2.req\endcsname{182.2}
\expandafter\def\csname ppt@rev.pol.rules.uiv.p2.replay-ppi-uniform@0.25.m2.req\endcsname{$+$182.2}
\expandafter\def\csname val@rev.pol.rules.uiv.p2.replay-ppi-uniform@0.25.m2.score\endcsname{0.830}
\expandafter\def\csname cnt@rev.pol.rules.uiv.p2.replay-ppi-uniform@0.25.m2.score\endcsname{1}
\expandafter\def\csname pct@rev.pol.rules.uiv.p2.replay-ppi-uniform@0.25.m2.score\endcsname{83.0}
\expandafter\def\csname ppt@rev.pol.rules.uiv.p2.replay-ppi-uniform@0.25.m2.score\endcsname{$+$83.0}
\expandafter\def\csname val@rev.pol.rules.uiv.p2.suspend-all.keep\endcsname{0.000}
\expandafter\def\csname cnt@rev.pol.rules.uiv.p2.suspend-all.keep\endcsname{0}
\expandafter\def\csname pct@rev.pol.rules.uiv.p2.suspend-all.keep\endcsname{0.0}
\expandafter\def\csname ppt@rev.pol.rules.uiv.p2.suspend-all.keep\endcsname{0.0}
\expandafter\def\csname val@rev.pol.rules.uiv.p2.suspend-all.req\endcsname{1.000}
\expandafter\def\csname cnt@rev.pol.rules.uiv.p2.suspend-all.req\endcsname{1}
\expandafter\def\csname pct@rev.pol.rules.uiv.p2.suspend-all.req\endcsname{100.0}
\expandafter\def\csname ppt@rev.pol.rules.uiv.p2.suspend-all.req\endcsname{$+$100.0}
\expandafter\def\csname val@rev.pol.rules.uiv.p2.suspend-all.score\endcsname{0.840}
\expandafter\def\csname cnt@rev.pol.rules.uiv.p2.suspend-all.score\endcsname{1}
\expandafter\def\csname pct@rev.pol.rules.uiv.p2.suspend-all.score\endcsname{84.0}
\expandafter\def\csname ppt@rev.pol.rules.uiv.p2.suspend-all.score\endcsname{$+$84.0}
\expandafter\def\csname val@rev.pol.rules.uiv.p2.uniform-fresh@0.25.frozen.keep\endcsname{0.718}
\expandafter\def\csname cnt@rev.pol.rules.uiv.p2.uniform-fresh@0.25.frozen.keep\endcsname{1}
\expandafter\def\csname pct@rev.pol.rules.uiv.p2.uniform-fresh@0.25.frozen.keep\endcsname{71.8}
\expandafter\def\csname ppt@rev.pol.rules.uiv.p2.uniform-fresh@0.25.frozen.keep\endcsname{$+$71.8}
\expandafter\def\csname val@rev.pol.rules.uiv.p2.uniform-fresh@0.25.frozen.req\endcsname{2.156}
\expandafter\def\csname cnt@rev.pol.rules.uiv.p2.uniform-fresh@0.25.frozen.req\endcsname{2}
\expandafter\def\csname pct@rev.pol.rules.uiv.p2.uniform-fresh@0.25.frozen.req\endcsname{215.6}
\expandafter\def\csname ppt@rev.pol.rules.uiv.p2.uniform-fresh@0.25.frozen.req\endcsname{$+$215.6}
\expandafter\def\csname val@rev.pol.rules.uiv.p2.uniform-fresh@0.25.frozen.score\endcsname{0.824}
\expandafter\def\csname cnt@rev.pol.rules.uiv.p2.uniform-fresh@0.25.frozen.score\endcsname{1}
\expandafter\def\csname pct@rev.pol.rules.uiv.p2.uniform-fresh@0.25.frozen.score\endcsname{82.4}
\expandafter\def\csname ppt@rev.pol.rules.uiv.p2.uniform-fresh@0.25.frozen.score\endcsname{$+$82.4}
\expandafter\def\csname val@rev.pol.rules.uiv.p2.uniform-fresh@0.25.lam0.5.keep\endcsname{0.396}
\expandafter\def\csname cnt@rev.pol.rules.uiv.p2.uniform-fresh@0.25.lam0.5.keep\endcsname{0}
\expandafter\def\csname pct@rev.pol.rules.uiv.p2.uniform-fresh@0.25.lam0.5.keep\endcsname{39.6}
\expandafter\def\csname ppt@rev.pol.rules.uiv.p2.uniform-fresh@0.25.lam0.5.keep\endcsname{$+$39.6}
\expandafter\def\csname val@rev.pol.rules.uiv.p2.uniform-fresh@0.25.lam0.5.req\endcsname{1.625}
\expandafter\def\csname cnt@rev.pol.rules.uiv.p2.uniform-fresh@0.25.lam0.5.req\endcsname{2}
\expandafter\def\csname pct@rev.pol.rules.uiv.p2.uniform-fresh@0.25.lam0.5.req\endcsname{162.5}
\expandafter\def\csname ppt@rev.pol.rules.uiv.p2.uniform-fresh@0.25.lam0.5.req\endcsname{$+$162.5}
\expandafter\def\csname val@rev.pol.rules.uiv.p2.uniform-fresh@0.25.lam0.5.score\endcsname{0.832}
\expandafter\def\csname cnt@rev.pol.rules.uiv.p2.uniform-fresh@0.25.lam0.5.score\endcsname{1}
\expandafter\def\csname pct@rev.pol.rules.uiv.p2.uniform-fresh@0.25.lam0.5.score\endcsname{83.2}
\expandafter\def\csname ppt@rev.pol.rules.uiv.p2.uniform-fresh@0.25.lam0.5.score\endcsname{$+$83.2}
\expandafter\def\csname val@rev.pol.rules.uiv.p2.uniform-fresh@0.25.lam1.keep\endcsname{0.396}
\expandafter\def\csname cnt@rev.pol.rules.uiv.p2.uniform-fresh@0.25.lam1.keep\endcsname{0}
\expandafter\def\csname pct@rev.pol.rules.uiv.p2.uniform-fresh@0.25.lam1.keep\endcsname{39.6}
\expandafter\def\csname ppt@rev.pol.rules.uiv.p2.uniform-fresh@0.25.lam1.keep\endcsname{$+$39.6}
\expandafter\def\csname val@rev.pol.rules.uiv.p2.uniform-fresh@0.25.lam1.req\endcsname{1.625}
\expandafter\def\csname cnt@rev.pol.rules.uiv.p2.uniform-fresh@0.25.lam1.req\endcsname{2}
\expandafter\def\csname pct@rev.pol.rules.uiv.p2.uniform-fresh@0.25.lam1.req\endcsname{162.5}
\expandafter\def\csname ppt@rev.pol.rules.uiv.p2.uniform-fresh@0.25.lam1.req\endcsname{$+$162.5}
\expandafter\def\csname val@rev.pol.rules.uiv.p2.uniform-fresh@0.25.lam1.score\endcsname{0.832}
\expandafter\def\csname cnt@rev.pol.rules.uiv.p2.uniform-fresh@0.25.lam1.score\endcsname{1}
\expandafter\def\csname pct@rev.pol.rules.uiv.p2.uniform-fresh@0.25.lam1.score\endcsname{83.2}
\expandafter\def\csname ppt@rev.pol.rules.uiv.p2.uniform-fresh@0.25.lam1.score\endcsname{$+$83.2}
\expandafter\def\csname val@rev.pol.rules.uiv.p2.uniform-fresh@0.25.m0.5.keep\endcsname{0.396}
\expandafter\def\csname cnt@rev.pol.rules.uiv.p2.uniform-fresh@0.25.m0.5.keep\endcsname{0}
\expandafter\def\csname pct@rev.pol.rules.uiv.p2.uniform-fresh@0.25.m0.5.keep\endcsname{39.6}
\expandafter\def\csname ppt@rev.pol.rules.uiv.p2.uniform-fresh@0.25.m0.5.keep\endcsname{$+$39.6}
\expandafter\def\csname val@rev.pol.rules.uiv.p2.uniform-fresh@0.25.m0.5.req\endcsname{1.625}
\expandafter\def\csname cnt@rev.pol.rules.uiv.p2.uniform-fresh@0.25.m0.5.req\endcsname{2}
\expandafter\def\csname pct@rev.pol.rules.uiv.p2.uniform-fresh@0.25.m0.5.req\endcsname{162.5}
\expandafter\def\csname ppt@rev.pol.rules.uiv.p2.uniform-fresh@0.25.m0.5.req\endcsname{$+$162.5}
\expandafter\def\csname val@rev.pol.rules.uiv.p2.uniform-fresh@0.25.m0.5.score\endcsname{0.832}
\expandafter\def\csname cnt@rev.pol.rules.uiv.p2.uniform-fresh@0.25.m0.5.score\endcsname{1}
\expandafter\def\csname pct@rev.pol.rules.uiv.p2.uniform-fresh@0.25.m0.5.score\endcsname{83.2}
\expandafter\def\csname ppt@rev.pol.rules.uiv.p2.uniform-fresh@0.25.m0.5.score\endcsname{$+$83.2}
\expandafter\def\csname val@rev.pol.rules.uiv.p2.uniform-fresh@0.25.m1.keep\endcsname{0.396}
\expandafter\def\csname cnt@rev.pol.rules.uiv.p2.uniform-fresh@0.25.m1.keep\endcsname{0}
\expandafter\def\csname pct@rev.pol.rules.uiv.p2.uniform-fresh@0.25.m1.keep\endcsname{39.6}
\expandafter\def\csname ppt@rev.pol.rules.uiv.p2.uniform-fresh@0.25.m1.keep\endcsname{$+$39.6}
\expandafter\def\csname val@rev.pol.rules.uiv.p2.uniform-fresh@0.25.m1.req\endcsname{1.625}
\expandafter\def\csname cnt@rev.pol.rules.uiv.p2.uniform-fresh@0.25.m1.req\endcsname{2}
\expandafter\def\csname pct@rev.pol.rules.uiv.p2.uniform-fresh@0.25.m1.req\endcsname{162.5}
\expandafter\def\csname ppt@rev.pol.rules.uiv.p2.uniform-fresh@0.25.m1.req\endcsname{$+$162.5}
\expandafter\def\csname val@rev.pol.rules.uiv.p2.uniform-fresh@0.25.m1.score\endcsname{0.832}
\expandafter\def\csname cnt@rev.pol.rules.uiv.p2.uniform-fresh@0.25.m1.score\endcsname{1}
\expandafter\def\csname pct@rev.pol.rules.uiv.p2.uniform-fresh@0.25.m1.score\endcsname{83.2}
\expandafter\def\csname ppt@rev.pol.rules.uiv.p2.uniform-fresh@0.25.m1.score\endcsname{$+$83.2}
\expandafter\def\csname val@rev.pol.rules.uiv.p2.uniform-fresh@0.25.m2.keep\endcsname{0.396}
\expandafter\def\csname cnt@rev.pol.rules.uiv.p2.uniform-fresh@0.25.m2.keep\endcsname{0}
\expandafter\def\csname pct@rev.pol.rules.uiv.p2.uniform-fresh@0.25.m2.keep\endcsname{39.6}
\expandafter\def\csname ppt@rev.pol.rules.uiv.p2.uniform-fresh@0.25.m2.keep\endcsname{$+$39.6}
\expandafter\def\csname val@rev.pol.rules.uiv.p2.uniform-fresh@0.25.m2.req\endcsname{1.625}
\expandafter\def\csname cnt@rev.pol.rules.uiv.p2.uniform-fresh@0.25.m2.req\endcsname{2}
\expandafter\def\csname pct@rev.pol.rules.uiv.p2.uniform-fresh@0.25.m2.req\endcsname{162.5}
\expandafter\def\csname ppt@rev.pol.rules.uiv.p2.uniform-fresh@0.25.m2.req\endcsname{$+$162.5}
\expandafter\def\csname val@rev.pol.rules.uiv.p2.uniform-fresh@0.25.m2.score\endcsname{0.832}
\expandafter\def\csname cnt@rev.pol.rules.uiv.p2.uniform-fresh@0.25.m2.score\endcsname{1}
\expandafter\def\csname pct@rev.pol.rules.uiv.p2.uniform-fresh@0.25.m2.score\endcsname{83.2}
\expandafter\def\csname ppt@rev.pol.rules.uiv.p2.uniform-fresh@0.25.m2.score\endcsname{$+$83.2}
\expandafter\def\csname val@rev.pol.signflip-mc\endcsname{200000}
\expandafter\def\csname cnt@rev.pol.signflip-mc\endcsname{200{,}000}
\expandafter\def\csname pct@rev.pol.signflip-mc\endcsname{20000000.0}
\expandafter\def\csname ppt@rev.pol.signflip-mc\endcsname{$+$20000000.0}
\expandafter\def\csname val@rev.pol.trig.disagree.minus-rcr-stratified.max\endcsname{0.029}
\expandafter\def\csname cnt@rev.pol.trig.disagree.minus-rcr-stratified.max\endcsname{0}
\expandafter\def\csname pct@rev.pol.trig.disagree.minus-rcr-stratified.max\endcsname{2.9}
\expandafter\def\csname ppt@rev.pol.trig.disagree.minus-rcr-stratified.max\endcsname{$+$2.9}
\expandafter\def\csname val@rev.pol.trig.disagree.minus-rcr-stratified.min\endcsname{-0.003}
\expandafter\def\csname cnt@rev.pol.trig.disagree.minus-rcr-stratified.min\endcsname{0}
\expandafter\def\csname pct@rev.pol.trig.disagree.minus-rcr-stratified.min\endcsname{$-$0.3}
\expandafter\def\csname ppt@rev.pol.trig.disagree.minus-rcr-stratified.min\endcsname{$-$0.3}
\expandafter\def\csname val@rev.pol.trig.disagree.minus-rcr-stratified.n-ci-above-0\endcsname{3}
\expandafter\def\csname cnt@rev.pol.trig.disagree.minus-rcr-stratified.n-ci-above-0\endcsname{3}
\expandafter\def\csname pct@rev.pol.trig.disagree.minus-rcr-stratified.n-ci-above-0\endcsname{300.0}
\expandafter\def\csname ppt@rev.pol.trig.disagree.minus-rcr-stratified.n-ci-above-0\endcsname{$+$300.0}
\expandafter\def\csname val@rev.pol.trig.disagree.minus-rcr-stratified.n-ci-below-0\endcsname{0}
\expandafter\def\csname cnt@rev.pol.trig.disagree.minus-rcr-stratified.n-ci-below-0\endcsname{0}
\expandafter\def\csname pct@rev.pol.trig.disagree.minus-rcr-stratified.n-ci-below-0\endcsname{0.0}
\expandafter\def\csname ppt@rev.pol.trig.disagree.minus-rcr-stratified.n-ci-below-0\endcsname{0.0}
\expandafter\def\csname val@rev.pol.trig.disagree.minus-replay-only.max\endcsname{0.000}
\expandafter\def\csname cnt@rev.pol.trig.disagree.minus-replay-only.max\endcsname{0}
\expandafter\def\csname pct@rev.pol.trig.disagree.minus-replay-only.max\endcsname{0.0}
\expandafter\def\csname ppt@rev.pol.trig.disagree.minus-replay-only.max\endcsname{0.0}
\expandafter\def\csname val@rev.pol.trig.disagree.minus-replay-only.min\endcsname{-0.008}
\expandafter\def\csname cnt@rev.pol.trig.disagree.minus-replay-only.min\endcsname{0}
\expandafter\def\csname pct@rev.pol.trig.disagree.minus-replay-only.min\endcsname{$-$0.8}
\expandafter\def\csname ppt@rev.pol.trig.disagree.minus-replay-only.min\endcsname{$-$0.8}
\expandafter\def\csname val@rev.pol.trig.disagree.minus-replay-only.n-ci-above-0\endcsname{0}
\expandafter\def\csname cnt@rev.pol.trig.disagree.minus-replay-only.n-ci-above-0\endcsname{0}
\expandafter\def\csname pct@rev.pol.trig.disagree.minus-replay-only.n-ci-above-0\endcsname{0.0}
\expandafter\def\csname ppt@rev.pol.trig.disagree.minus-replay-only.n-ci-above-0\endcsname{0.0}
\expandafter\def\csname val@rev.pol.trig.disagree.minus-replay-only.n-ci-below-0\endcsname{1}
\expandafter\def\csname cnt@rev.pol.trig.disagree.minus-replay-only.n-ci-below-0\endcsname{1}
\expandafter\def\csname pct@rev.pol.trig.disagree.minus-replay-only.n-ci-below-0\endcsname{100.0}
\expandafter\def\csname ppt@rev.pol.trig.disagree.minus-replay-only.n-ci-below-0\endcsname{$+$100.0}
\expandafter\def\csname val@rev.pol.trig.disagree.minus-replay-ppi-uniform.max\endcsname{0.033}
\expandafter\def\csname cnt@rev.pol.trig.disagree.minus-replay-ppi-uniform.max\endcsname{0}
\expandafter\def\csname pct@rev.pol.trig.disagree.minus-replay-ppi-uniform.max\endcsname{3.3}
\expandafter\def\csname ppt@rev.pol.trig.disagree.minus-replay-ppi-uniform.max\endcsname{$+$3.3}
\expandafter\def\csname val@rev.pol.trig.disagree.minus-replay-ppi-uniform.min\endcsname{-0.002}
\expandafter\def\csname cnt@rev.pol.trig.disagree.minus-replay-ppi-uniform.min\endcsname{0}
\expandafter\def\csname pct@rev.pol.trig.disagree.minus-replay-ppi-uniform.min\endcsname{$-$0.2}
\expandafter\def\csname ppt@rev.pol.trig.disagree.minus-replay-ppi-uniform.min\endcsname{$-$0.2}
\expandafter\def\csname val@rev.pol.trig.disagree.minus-replay-ppi-uniform.n-ci-above-0\endcsname{3}
\expandafter\def\csname cnt@rev.pol.trig.disagree.minus-replay-ppi-uniform.n-ci-above-0\endcsname{3}
\expandafter\def\csname pct@rev.pol.trig.disagree.minus-replay-ppi-uniform.n-ci-above-0\endcsname{300.0}
\expandafter\def\csname ppt@rev.pol.trig.disagree.minus-replay-ppi-uniform.n-ci-above-0\endcsname{$+$300.0}
\expandafter\def\csname val@rev.pol.trig.disagree.minus-replay-ppi-uniform.n-ci-below-0\endcsname{0}
\expandafter\def\csname cnt@rev.pol.trig.disagree.minus-replay-ppi-uniform.n-ci-below-0\endcsname{0}
\expandafter\def\csname pct@rev.pol.trig.disagree.minus-replay-ppi-uniform.n-ci-below-0\endcsname{0.0}
\expandafter\def\csname ppt@rev.pol.trig.disagree.minus-replay-ppi-uniform.n-ci-below-0\endcsname{0.0}
\expandafter\def\csname val@rev.pol.trig.disagree.n-doubt-total\endcsname{2}
\expandafter\def\csname cnt@rev.pol.trig.disagree.n-doubt-total\endcsname{2}
\expandafter\def\csname pct@rev.pol.trig.disagree.n-doubt-total\endcsname{200.0}
\expandafter\def\csname ppt@rev.pol.trig.disagree.n-doubt-total\endcsname{$+$200.0}
\expandafter\def\csname val@rev.pol.trig.disagree.sspro.p1.cost\endcsname{96.000}
\expandafter\def\csname cnt@rev.pol.trig.disagree.sspro.p1.cost\endcsname{96}
\expandafter\def\csname pct@rev.pol.trig.disagree.sspro.p1.cost\endcsname{9600.0}
\expandafter\def\csname ppt@rev.pol.trig.disagree.sspro.p1.cost\endcsname{$+$9600.0}
\expandafter\def\csname val@rev.pol.trig.disagree.sspro.p1.keep\endcsname{1.000}
\expandafter\def\csname cnt@rev.pol.trig.disagree.sspro.p1.keep\endcsname{1}
\expandafter\def\csname pct@rev.pol.trig.disagree.sspro.p1.keep\endcsname{100.0}
\expandafter\def\csname ppt@rev.pol.trig.disagree.sspro.p1.keep\endcsname{$+$100.0}
\expandafter\def\csname val@rev.pol.trig.disagree.sspro.p1.minus-rcr-stratified.ci95.0\endcsname{0.000}
\expandafter\def\csname cnt@rev.pol.trig.disagree.sspro.p1.minus-rcr-stratified.ci95.0\endcsname{0}
\expandafter\def\csname pct@rev.pol.trig.disagree.sspro.p1.minus-rcr-stratified.ci95.0\endcsname{0.0}
\expandafter\def\csname ppt@rev.pol.trig.disagree.sspro.p1.minus-rcr-stratified.ci95.0\endcsname{0.0}
\expandafter\def\csname val@rev.pol.trig.disagree.sspro.p1.minus-rcr-stratified.ci95.1\endcsname{0.007}
\expandafter\def\csname cnt@rev.pol.trig.disagree.sspro.p1.minus-rcr-stratified.ci95.1\endcsname{0}
\expandafter\def\csname pct@rev.pol.trig.disagree.sspro.p1.minus-rcr-stratified.ci95.1\endcsname{0.7}
\expandafter\def\csname ppt@rev.pol.trig.disagree.sspro.p1.minus-rcr-stratified.ci95.1\endcsname{$+$0.7}
\expandafter\def\csname val@rev.pol.trig.disagree.sspro.p1.minus-rcr-stratified.mean\endcsname{0.004}
\expandafter\def\csname cnt@rev.pol.trig.disagree.sspro.p1.minus-rcr-stratified.mean\endcsname{0}
\expandafter\def\csname pct@rev.pol.trig.disagree.sspro.p1.minus-rcr-stratified.mean\endcsname{0.4}
\expandafter\def\csname ppt@rev.pol.trig.disagree.sspro.p1.minus-rcr-stratified.mean\endcsname{$+$0.4}
\expandafter\def\csname val@rev.pol.trig.disagree.sspro.p1.minus-replay-only.ci95.0\endcsname{0.000}
\expandafter\def\csname cnt@rev.pol.trig.disagree.sspro.p1.minus-replay-only.ci95.0\endcsname{0}
\expandafter\def\csname pct@rev.pol.trig.disagree.sspro.p1.minus-replay-only.ci95.0\endcsname{0.0}
\expandafter\def\csname ppt@rev.pol.trig.disagree.sspro.p1.minus-replay-only.ci95.0\endcsname{0.0}
\expandafter\def\csname val@rev.pol.trig.disagree.sspro.p1.minus-replay-only.ci95.1\endcsname{0.000}
\expandafter\def\csname cnt@rev.pol.trig.disagree.sspro.p1.minus-replay-only.ci95.1\endcsname{0}
\expandafter\def\csname pct@rev.pol.trig.disagree.sspro.p1.minus-replay-only.ci95.1\endcsname{0.0}
\expandafter\def\csname ppt@rev.pol.trig.disagree.sspro.p1.minus-replay-only.ci95.1\endcsname{0.0}
\expandafter\def\csname val@rev.pol.trig.disagree.sspro.p1.minus-replay-only.mean\endcsname{0.000}
\expandafter\def\csname cnt@rev.pol.trig.disagree.sspro.p1.minus-replay-only.mean\endcsname{0}
\expandafter\def\csname pct@rev.pol.trig.disagree.sspro.p1.minus-replay-only.mean\endcsname{0.0}
\expandafter\def\csname ppt@rev.pol.trig.disagree.sspro.p1.minus-replay-only.mean\endcsname{0.0}
\expandafter\def\csname val@rev.pol.trig.disagree.sspro.p1.minus-replay-ppi-uniform.ci95.0\endcsname{0.001}
\expandafter\def\csname cnt@rev.pol.trig.disagree.sspro.p1.minus-replay-ppi-uniform.ci95.0\endcsname{0}
\expandafter\def\csname pct@rev.pol.trig.disagree.sspro.p1.minus-replay-ppi-uniform.ci95.0\endcsname{0.1}
\expandafter\def\csname ppt@rev.pol.trig.disagree.sspro.p1.minus-replay-ppi-uniform.ci95.0\endcsname{$+$0.1}
\expandafter\def\csname val@rev.pol.trig.disagree.sspro.p1.minus-replay-ppi-uniform.ci95.1\endcsname{0.014}
\expandafter\def\csname cnt@rev.pol.trig.disagree.sspro.p1.minus-replay-ppi-uniform.ci95.1\endcsname{0}
\expandafter\def\csname pct@rev.pol.trig.disagree.sspro.p1.minus-replay-ppi-uniform.ci95.1\endcsname{1.4}
\expandafter\def\csname ppt@rev.pol.trig.disagree.sspro.p1.minus-replay-ppi-uniform.ci95.1\endcsname{$+$1.4}
\expandafter\def\csname val@rev.pol.trig.disagree.sspro.p1.minus-replay-ppi-uniform.mean\endcsname{0.007}
\expandafter\def\csname cnt@rev.pol.trig.disagree.sspro.p1.minus-replay-ppi-uniform.mean\endcsname{0}
\expandafter\def\csname pct@rev.pol.trig.disagree.sspro.p1.minus-replay-ppi-uniform.mean\endcsname{0.7}
\expandafter\def\csname ppt@rev.pol.trig.disagree.sspro.p1.minus-replay-ppi-uniform.mean\endcsname{$+$0.7}
\expandafter\def\csname val@rev.pol.trig.disagree.sspro.p1.n-doubt\endcsname{0}
\expandafter\def\csname cnt@rev.pol.trig.disagree.sspro.p1.n-doubt\endcsname{0}
\expandafter\def\csname pct@rev.pol.trig.disagree.sspro.p1.n-doubt\endcsname{0.0}
\expandafter\def\csname ppt@rev.pol.trig.disagree.sspro.p1.n-doubt\endcsname{0.0}
\expandafter\def\csname val@rev.pol.trig.disagree.sspro.p1.score\endcsname{0.859}
\expandafter\def\csname cnt@rev.pol.trig.disagree.sspro.p1.score\endcsname{1}
\expandafter\def\csname pct@rev.pol.trig.disagree.sspro.p1.score\endcsname{85.9}
\expandafter\def\csname ppt@rev.pol.trig.disagree.sspro.p1.score\endcsname{$+$85.9}
\expandafter\def\csname val@rev.pol.trig.disagree.sspro.p1.sd\endcsname{0.000}
\expandafter\def\csname cnt@rev.pol.trig.disagree.sspro.p1.sd\endcsname{0}
\expandafter\def\csname pct@rev.pol.trig.disagree.sspro.p1.sd\endcsname{0.0}
\expandafter\def\csname ppt@rev.pol.trig.disagree.sspro.p1.sd\endcsname{0.0}
\expandafter\def\csname val@rev.pol.trig.disagree.sspro.p2.cost\endcsname{96.000}
\expandafter\def\csname cnt@rev.pol.trig.disagree.sspro.p2.cost\endcsname{96}
\expandafter\def\csname pct@rev.pol.trig.disagree.sspro.p2.cost\endcsname{9600.0}
\expandafter\def\csname ppt@rev.pol.trig.disagree.sspro.p2.cost\endcsname{$+$9600.0}
\expandafter\def\csname val@rev.pol.trig.disagree.sspro.p2.keep\endcsname{1.000}
\expandafter\def\csname cnt@rev.pol.trig.disagree.sspro.p2.keep\endcsname{1}
\expandafter\def\csname pct@rev.pol.trig.disagree.sspro.p2.keep\endcsname{100.0}
\expandafter\def\csname ppt@rev.pol.trig.disagree.sspro.p2.keep\endcsname{$+$100.0}
\expandafter\def\csname val@rev.pol.trig.disagree.sspro.p2.minus-rcr-stratified.ci95.0\endcsname{0.009}
\expandafter\def\csname cnt@rev.pol.trig.disagree.sspro.p2.minus-rcr-stratified.ci95.0\endcsname{0}
\expandafter\def\csname pct@rev.pol.trig.disagree.sspro.p2.minus-rcr-stratified.ci95.0\endcsname{0.9}
\expandafter\def\csname ppt@rev.pol.trig.disagree.sspro.p2.minus-rcr-stratified.ci95.0\endcsname{$+$0.9}
\expandafter\def\csname val@rev.pol.trig.disagree.sspro.p2.minus-rcr-stratified.ci95.1\endcsname{0.023}
\expandafter\def\csname cnt@rev.pol.trig.disagree.sspro.p2.minus-rcr-stratified.ci95.1\endcsname{0}
\expandafter\def\csname pct@rev.pol.trig.disagree.sspro.p2.minus-rcr-stratified.ci95.1\endcsname{2.3}
\expandafter\def\csname ppt@rev.pol.trig.disagree.sspro.p2.minus-rcr-stratified.ci95.1\endcsname{$+$2.3}
\expandafter\def\csname val@rev.pol.trig.disagree.sspro.p2.minus-rcr-stratified.mean\endcsname{0.016}
\expandafter\def\csname cnt@rev.pol.trig.disagree.sspro.p2.minus-rcr-stratified.mean\endcsname{0}
\expandafter\def\csname pct@rev.pol.trig.disagree.sspro.p2.minus-rcr-stratified.mean\endcsname{1.6}
\expandafter\def\csname ppt@rev.pol.trig.disagree.sspro.p2.minus-rcr-stratified.mean\endcsname{$+$1.6}
\expandafter\def\csname val@rev.pol.trig.disagree.sspro.p2.minus-replay-only.ci95.0\endcsname{0.000}
\expandafter\def\csname cnt@rev.pol.trig.disagree.sspro.p2.minus-replay-only.ci95.0\endcsname{0}
\expandafter\def\csname pct@rev.pol.trig.disagree.sspro.p2.minus-replay-only.ci95.0\endcsname{0.0}
\expandafter\def\csname ppt@rev.pol.trig.disagree.sspro.p2.minus-replay-only.ci95.0\endcsname{0.0}
\expandafter\def\csname val@rev.pol.trig.disagree.sspro.p2.minus-replay-only.ci95.1\endcsname{0.000}
\expandafter\def\csname cnt@rev.pol.trig.disagree.sspro.p2.minus-replay-only.ci95.1\endcsname{0}
\expandafter\def\csname pct@rev.pol.trig.disagree.sspro.p2.minus-replay-only.ci95.1\endcsname{0.0}
\expandafter\def\csname ppt@rev.pol.trig.disagree.sspro.p2.minus-replay-only.ci95.1\endcsname{0.0}
\expandafter\def\csname val@rev.pol.trig.disagree.sspro.p2.minus-replay-only.mean\endcsname{0.000}
\expandafter\def\csname cnt@rev.pol.trig.disagree.sspro.p2.minus-replay-only.mean\endcsname{0}
\expandafter\def\csname pct@rev.pol.trig.disagree.sspro.p2.minus-replay-only.mean\endcsname{0.0}
\expandafter\def\csname ppt@rev.pol.trig.disagree.sspro.p2.minus-replay-only.mean\endcsname{0.0}
\expandafter\def\csname val@rev.pol.trig.disagree.sspro.p2.minus-replay-ppi-uniform.ci95.0\endcsname{0.009}
\expandafter\def\csname cnt@rev.pol.trig.disagree.sspro.p2.minus-replay-ppi-uniform.ci95.0\endcsname{0}
\expandafter\def\csname pct@rev.pol.trig.disagree.sspro.p2.minus-replay-ppi-uniform.ci95.0\endcsname{0.9}
\expandafter\def\csname ppt@rev.pol.trig.disagree.sspro.p2.minus-replay-ppi-uniform.ci95.0\endcsname{$+$0.9}
\expandafter\def\csname val@rev.pol.trig.disagree.sspro.p2.minus-replay-ppi-uniform.ci95.1\endcsname{0.023}
\expandafter\def\csname cnt@rev.pol.trig.disagree.sspro.p2.minus-replay-ppi-uniform.ci95.1\endcsname{0}
\expandafter\def\csname pct@rev.pol.trig.disagree.sspro.p2.minus-replay-ppi-uniform.ci95.1\endcsname{2.3}
\expandafter\def\csname ppt@rev.pol.trig.disagree.sspro.p2.minus-replay-ppi-uniform.ci95.1\endcsname{$+$2.3}
\expandafter\def\csname val@rev.pol.trig.disagree.sspro.p2.minus-replay-ppi-uniform.mean\endcsname{0.016}
\expandafter\def\csname cnt@rev.pol.trig.disagree.sspro.p2.minus-replay-ppi-uniform.mean\endcsname{0}
\expandafter\def\csname pct@rev.pol.trig.disagree.sspro.p2.minus-replay-ppi-uniform.mean\endcsname{1.6}
\expandafter\def\csname ppt@rev.pol.trig.disagree.sspro.p2.minus-replay-ppi-uniform.mean\endcsname{$+$1.6}
\expandafter\def\csname val@rev.pol.trig.disagree.sspro.p2.n-doubt\endcsname{0}
\expandafter\def\csname cnt@rev.pol.trig.disagree.sspro.p2.n-doubt\endcsname{0}
\expandafter\def\csname pct@rev.pol.trig.disagree.sspro.p2.n-doubt\endcsname{0.0}
\expandafter\def\csname ppt@rev.pol.trig.disagree.sspro.p2.n-doubt\endcsname{0.0}
\expandafter\def\csname val@rev.pol.trig.disagree.sspro.p2.score\endcsname{0.906}
\expandafter\def\csname cnt@rev.pol.trig.disagree.sspro.p2.score\endcsname{1}
\expandafter\def\csname pct@rev.pol.trig.disagree.sspro.p2.score\endcsname{90.6}
\expandafter\def\csname ppt@rev.pol.trig.disagree.sspro.p2.score\endcsname{$+$90.6}
\expandafter\def\csname val@rev.pol.trig.disagree.sspro.p2.sd\endcsname{0.000}
\expandafter\def\csname cnt@rev.pol.trig.disagree.sspro.p2.sd\endcsname{0}
\expandafter\def\csname pct@rev.pol.trig.disagree.sspro.p2.sd\endcsname{0.0}
\expandafter\def\csname ppt@rev.pol.trig.disagree.sspro.p2.sd\endcsname{0.0}
\expandafter\def\csname val@rev.pol.trig.disagree.uiv.p1.cost\endcsname{192.000}
\expandafter\def\csname cnt@rev.pol.trig.disagree.uiv.p1.cost\endcsname{192}
\expandafter\def\csname pct@rev.pol.trig.disagree.uiv.p1.cost\endcsname{19200.0}
\expandafter\def\csname ppt@rev.pol.trig.disagree.uiv.p1.cost\endcsname{$+$19200.0}
\expandafter\def\csname val@rev.pol.trig.disagree.uiv.p1.keep\endcsname{0.000}
\expandafter\def\csname cnt@rev.pol.trig.disagree.uiv.p1.keep\endcsname{0}
\expandafter\def\csname pct@rev.pol.trig.disagree.uiv.p1.keep\endcsname{0.0}
\expandafter\def\csname ppt@rev.pol.trig.disagree.uiv.p1.keep\endcsname{0.0}
\expandafter\def\csname val@rev.pol.trig.disagree.uiv.p1.minus-rcr-stratified.ci95.0\endcsname{0.014}
\expandafter\def\csname cnt@rev.pol.trig.disagree.uiv.p1.minus-rcr-stratified.ci95.0\endcsname{0}
\expandafter\def\csname pct@rev.pol.trig.disagree.uiv.p1.minus-rcr-stratified.ci95.0\endcsname{1.4}
\expandafter\def\csname ppt@rev.pol.trig.disagree.uiv.p1.minus-rcr-stratified.ci95.0\endcsname{$+$1.4}
\expandafter\def\csname val@rev.pol.trig.disagree.uiv.p1.minus-rcr-stratified.ci95.1\endcsname{0.044}
\expandafter\def\csname cnt@rev.pol.trig.disagree.uiv.p1.minus-rcr-stratified.ci95.1\endcsname{0}
\expandafter\def\csname pct@rev.pol.trig.disagree.uiv.p1.minus-rcr-stratified.ci95.1\endcsname{4.4}
\expandafter\def\csname ppt@rev.pol.trig.disagree.uiv.p1.minus-rcr-stratified.ci95.1\endcsname{$+$4.4}
\expandafter\def\csname val@rev.pol.trig.disagree.uiv.p1.minus-rcr-stratified.mean\endcsname{0.029}
\expandafter\def\csname cnt@rev.pol.trig.disagree.uiv.p1.minus-rcr-stratified.mean\endcsname{0}
\expandafter\def\csname pct@rev.pol.trig.disagree.uiv.p1.minus-rcr-stratified.mean\endcsname{2.9}
\expandafter\def\csname ppt@rev.pol.trig.disagree.uiv.p1.minus-rcr-stratified.mean\endcsname{$+$2.9}
\expandafter\def\csname val@rev.pol.trig.disagree.uiv.p1.minus-replay-only.ci95.0\endcsname{0.000}
\expandafter\def\csname cnt@rev.pol.trig.disagree.uiv.p1.minus-replay-only.ci95.0\endcsname{0}
\expandafter\def\csname pct@rev.pol.trig.disagree.uiv.p1.minus-replay-only.ci95.0\endcsname{0.0}
\expandafter\def\csname ppt@rev.pol.trig.disagree.uiv.p1.minus-replay-only.ci95.0\endcsname{0.0}
\expandafter\def\csname val@rev.pol.trig.disagree.uiv.p1.minus-replay-only.ci95.1\endcsname{0.000}
\expandafter\def\csname cnt@rev.pol.trig.disagree.uiv.p1.minus-replay-only.ci95.1\endcsname{0}
\expandafter\def\csname pct@rev.pol.trig.disagree.uiv.p1.minus-replay-only.ci95.1\endcsname{0.0}
\expandafter\def\csname ppt@rev.pol.trig.disagree.uiv.p1.minus-replay-only.ci95.1\endcsname{0.0}
\expandafter\def\csname val@rev.pol.trig.disagree.uiv.p1.minus-replay-only.mean\endcsname{0.000}
\expandafter\def\csname cnt@rev.pol.trig.disagree.uiv.p1.minus-replay-only.mean\endcsname{0}
\expandafter\def\csname pct@rev.pol.trig.disagree.uiv.p1.minus-replay-only.mean\endcsname{0.0}
\expandafter\def\csname ppt@rev.pol.trig.disagree.uiv.p1.minus-replay-only.mean\endcsname{0.0}
\expandafter\def\csname val@rev.pol.trig.disagree.uiv.p1.minus-replay-ppi-uniform.ci95.0\endcsname{0.016}
\expandafter\def\csname cnt@rev.pol.trig.disagree.uiv.p1.minus-replay-ppi-uniform.ci95.0\endcsname{0}
\expandafter\def\csname pct@rev.pol.trig.disagree.uiv.p1.minus-replay-ppi-uniform.ci95.0\endcsname{1.6}
\expandafter\def\csname ppt@rev.pol.trig.disagree.uiv.p1.minus-replay-ppi-uniform.ci95.0\endcsname{$+$1.6}
\expandafter\def\csname val@rev.pol.trig.disagree.uiv.p1.minus-replay-ppi-uniform.ci95.1\endcsname{0.051}
\expandafter\def\csname cnt@rev.pol.trig.disagree.uiv.p1.minus-replay-ppi-uniform.ci95.1\endcsname{0}
\expandafter\def\csname pct@rev.pol.trig.disagree.uiv.p1.minus-replay-ppi-uniform.ci95.1\endcsname{5.1}
\expandafter\def\csname ppt@rev.pol.trig.disagree.uiv.p1.minus-replay-ppi-uniform.ci95.1\endcsname{$+$5.1}
\expandafter\def\csname val@rev.pol.trig.disagree.uiv.p1.minus-replay-ppi-uniform.mean\endcsname{0.033}
\expandafter\def\csname cnt@rev.pol.trig.disagree.uiv.p1.minus-replay-ppi-uniform.mean\endcsname{0}
\expandafter\def\csname pct@rev.pol.trig.disagree.uiv.p1.minus-replay-ppi-uniform.mean\endcsname{3.3}
\expandafter\def\csname ppt@rev.pol.trig.disagree.uiv.p1.minus-replay-ppi-uniform.mean\endcsname{$+$3.3}
\expandafter\def\csname val@rev.pol.trig.disagree.uiv.p1.n-doubt\endcsname{0}
\expandafter\def\csname cnt@rev.pol.trig.disagree.uiv.p1.n-doubt\endcsname{0}
\expandafter\def\csname pct@rev.pol.trig.disagree.uiv.p1.n-doubt\endcsname{0.0}
\expandafter\def\csname ppt@rev.pol.trig.disagree.uiv.p1.n-doubt\endcsname{0.0}
\expandafter\def\csname val@rev.pol.trig.disagree.uiv.p1.score\endcsname{0.833}
\expandafter\def\csname cnt@rev.pol.trig.disagree.uiv.p1.score\endcsname{1}
\expandafter\def\csname pct@rev.pol.trig.disagree.uiv.p1.score\endcsname{83.3}
\expandafter\def\csname ppt@rev.pol.trig.disagree.uiv.p1.score\endcsname{$+$83.3}
\expandafter\def\csname val@rev.pol.trig.disagree.uiv.p1.sd\endcsname{0.000}
\expandafter\def\csname cnt@rev.pol.trig.disagree.uiv.p1.sd\endcsname{0}
\expandafter\def\csname pct@rev.pol.trig.disagree.uiv.p1.sd\endcsname{0.0}
\expandafter\def\csname ppt@rev.pol.trig.disagree.uiv.p1.sd\endcsname{0.0}
\expandafter\def\csname val@rev.pol.trig.disagree.uiv.p2.cost\endcsname{252.000}
\expandafter\def\csname cnt@rev.pol.trig.disagree.uiv.p2.cost\endcsname{252}
\expandafter\def\csname pct@rev.pol.trig.disagree.uiv.p2.cost\endcsname{25200.0}
\expandafter\def\csname ppt@rev.pol.trig.disagree.uiv.p2.cost\endcsname{$+$25200.0}
\expandafter\def\csname val@rev.pol.trig.disagree.uiv.p2.keep\endcsname{0.567}
\expandafter\def\csname cnt@rev.pol.trig.disagree.uiv.p2.keep\endcsname{1}
\expandafter\def\csname pct@rev.pol.trig.disagree.uiv.p2.keep\endcsname{56.7}
\expandafter\def\csname ppt@rev.pol.trig.disagree.uiv.p2.keep\endcsname{$+$56.7}
\expandafter\def\csname val@rev.pol.trig.disagree.uiv.p2.minus-rcr-stratified.ci95.0\endcsname{-0.008}
\expandafter\def\csname cnt@rev.pol.trig.disagree.uiv.p2.minus-rcr-stratified.ci95.0\endcsname{0}
\expandafter\def\csname pct@rev.pol.trig.disagree.uiv.p2.minus-rcr-stratified.ci95.0\endcsname{$-$0.8}
\expandafter\def\csname ppt@rev.pol.trig.disagree.uiv.p2.minus-rcr-stratified.ci95.0\endcsname{$-$0.8}
\expandafter\def\csname val@rev.pol.trig.disagree.uiv.p2.minus-rcr-stratified.ci95.1\endcsname{0.002}
\expandafter\def\csname cnt@rev.pol.trig.disagree.uiv.p2.minus-rcr-stratified.ci95.1\endcsname{0}
\expandafter\def\csname pct@rev.pol.trig.disagree.uiv.p2.minus-rcr-stratified.ci95.1\endcsname{0.2}
\expandafter\def\csname ppt@rev.pol.trig.disagree.uiv.p2.minus-rcr-stratified.ci95.1\endcsname{$+$0.2}
\expandafter\def\csname val@rev.pol.trig.disagree.uiv.p2.minus-rcr-stratified.mean\endcsname{-0.003}
\expandafter\def\csname cnt@rev.pol.trig.disagree.uiv.p2.minus-rcr-stratified.mean\endcsname{0}
\expandafter\def\csname pct@rev.pol.trig.disagree.uiv.p2.minus-rcr-stratified.mean\endcsname{$-$0.3}
\expandafter\def\csname ppt@rev.pol.trig.disagree.uiv.p2.minus-rcr-stratified.mean\endcsname{$-$0.3}
\expandafter\def\csname val@rev.pol.trig.disagree.uiv.p2.minus-replay-only.ci95.0\endcsname{-0.015}
\expandafter\def\csname cnt@rev.pol.trig.disagree.uiv.p2.minus-replay-only.ci95.0\endcsname{0}
\expandafter\def\csname pct@rev.pol.trig.disagree.uiv.p2.minus-replay-only.ci95.0\endcsname{$-$1.5}
\expandafter\def\csname ppt@rev.pol.trig.disagree.uiv.p2.minus-replay-only.ci95.0\endcsname{$-$1.5}
\expandafter\def\csname val@rev.pol.trig.disagree.uiv.p2.minus-replay-only.ci95.1\endcsname{-0.001}
\expandafter\def\csname cnt@rev.pol.trig.disagree.uiv.p2.minus-replay-only.ci95.1\endcsname{0}
\expandafter\def\csname pct@rev.pol.trig.disagree.uiv.p2.minus-replay-only.ci95.1\endcsname{$-$0.1}
\expandafter\def\csname ppt@rev.pol.trig.disagree.uiv.p2.minus-replay-only.ci95.1\endcsname{$-$0.1}
\expandafter\def\csname val@rev.pol.trig.disagree.uiv.p2.minus-replay-only.mean\endcsname{-0.008}
\expandafter\def\csname cnt@rev.pol.trig.disagree.uiv.p2.minus-replay-only.mean\endcsname{0}
\expandafter\def\csname pct@rev.pol.trig.disagree.uiv.p2.minus-replay-only.mean\endcsname{$-$0.8}
\expandafter\def\csname ppt@rev.pol.trig.disagree.uiv.p2.minus-replay-only.mean\endcsname{$-$0.8}
\expandafter\def\csname val@rev.pol.trig.disagree.uiv.p2.minus-replay-ppi-uniform.ci95.0\endcsname{-0.006}
\expandafter\def\csname cnt@rev.pol.trig.disagree.uiv.p2.minus-replay-ppi-uniform.ci95.0\endcsname{0}
\expandafter\def\csname pct@rev.pol.trig.disagree.uiv.p2.minus-replay-ppi-uniform.ci95.0\endcsname{$-$0.6}
\expandafter\def\csname ppt@rev.pol.trig.disagree.uiv.p2.minus-replay-ppi-uniform.ci95.0\endcsname{$-$0.6}
\expandafter\def\csname val@rev.pol.trig.disagree.uiv.p2.minus-replay-ppi-uniform.ci95.1\endcsname{0.003}
\expandafter\def\csname cnt@rev.pol.trig.disagree.uiv.p2.minus-replay-ppi-uniform.ci95.1\endcsname{0}
\expandafter\def\csname pct@rev.pol.trig.disagree.uiv.p2.minus-replay-ppi-uniform.ci95.1\endcsname{0.3}
\expandafter\def\csname ppt@rev.pol.trig.disagree.uiv.p2.minus-replay-ppi-uniform.ci95.1\endcsname{$+$0.3}
\expandafter\def\csname val@rev.pol.trig.disagree.uiv.p2.minus-replay-ppi-uniform.mean\endcsname{-0.002}
\expandafter\def\csname cnt@rev.pol.trig.disagree.uiv.p2.minus-replay-ppi-uniform.mean\endcsname{0}
\expandafter\def\csname pct@rev.pol.trig.disagree.uiv.p2.minus-replay-ppi-uniform.mean\endcsname{$-$0.2}
\expandafter\def\csname ppt@rev.pol.trig.disagree.uiv.p2.minus-replay-ppi-uniform.mean\endcsname{$-$0.2}
\expandafter\def\csname val@rev.pol.trig.disagree.uiv.p2.n-doubt\endcsname{2}
\expandafter\def\csname cnt@rev.pol.trig.disagree.uiv.p2.n-doubt\endcsname{2}
\expandafter\def\csname pct@rev.pol.trig.disagree.uiv.p2.n-doubt\endcsname{200.0}
\expandafter\def\csname ppt@rev.pol.trig.disagree.uiv.p2.n-doubt\endcsname{$+$200.0}
\expandafter\def\csname val@rev.pol.trig.disagree.uiv.p2.score\endcsname{0.826}
\expandafter\def\csname cnt@rev.pol.trig.disagree.uiv.p2.score\endcsname{1}
\expandafter\def\csname pct@rev.pol.trig.disagree.uiv.p2.score\endcsname{82.6}
\expandafter\def\csname ppt@rev.pol.trig.disagree.uiv.p2.score\endcsname{$+$82.6}
\expandafter\def\csname val@rev.pol.trig.disagree.uiv.p2.sd\endcsname{0.008}
\expandafter\def\csname cnt@rev.pol.trig.disagree.uiv.p2.sd\endcsname{0}
\expandafter\def\csname pct@rev.pol.trig.disagree.uiv.p2.sd\endcsname{0.8}
\expandafter\def\csname ppt@rev.pol.trig.disagree.uiv.p2.sd\endcsname{$+$0.8}
\expandafter\def\csname val@rev.pol.trig.n-skill-settings\endcsname{16}
\expandafter\def\csname cnt@rev.pol.trig.n-skill-settings\endcsname{16}
\expandafter\def\csname pct@rev.pol.trig.n-skill-settings\endcsname{1600.0}
\expandafter\def\csname ppt@rev.pol.trig.n-skill-settings\endcsname{$+$1600.0}
\expandafter\def\csname val@rev.pol.trig.se.minus-rcr-stratified.max\endcsname{0.004}
\expandafter\def\csname cnt@rev.pol.trig.se.minus-rcr-stratified.max\endcsname{0}
\expandafter\def\csname pct@rev.pol.trig.se.minus-rcr-stratified.max\endcsname{0.4}
\expandafter\def\csname ppt@rev.pol.trig.se.minus-rcr-stratified.max\endcsname{$+$0.4}
\expandafter\def\csname val@rev.pol.trig.se.minus-rcr-stratified.min\endcsname{-0.003}
\expandafter\def\csname cnt@rev.pol.trig.se.minus-rcr-stratified.min\endcsname{0}
\expandafter\def\csname pct@rev.pol.trig.se.minus-rcr-stratified.min\endcsname{$-$0.3}
\expandafter\def\csname ppt@rev.pol.trig.se.minus-rcr-stratified.min\endcsname{$-$0.3}
\expandafter\def\csname val@rev.pol.trig.se.minus-rcr-stratified.n-ci-above-0\endcsname{2}
\expandafter\def\csname cnt@rev.pol.trig.se.minus-rcr-stratified.n-ci-above-0\endcsname{2}
\expandafter\def\csname pct@rev.pol.trig.se.minus-rcr-stratified.n-ci-above-0\endcsname{200.0}
\expandafter\def\csname ppt@rev.pol.trig.se.minus-rcr-stratified.n-ci-above-0\endcsname{$+$200.0}
\expandafter\def\csname val@rev.pol.trig.se.minus-rcr-stratified.n-ci-below-0\endcsname{2}
\expandafter\def\csname cnt@rev.pol.trig.se.minus-rcr-stratified.n-ci-below-0\endcsname{2}
\expandafter\def\csname pct@rev.pol.trig.se.minus-rcr-stratified.n-ci-below-0\endcsname{200.0}
\expandafter\def\csname ppt@rev.pol.trig.se.minus-rcr-stratified.n-ci-below-0\endcsname{$+$200.0}
\expandafter\def\csname val@rev.pol.trig.se.minus-replay-only.max\endcsname{-0.006}
\expandafter\def\csname cnt@rev.pol.trig.se.minus-replay-only.max\endcsname{0}
\expandafter\def\csname pct@rev.pol.trig.se.minus-replay-only.max\endcsname{$-$0.6}
\expandafter\def\csname ppt@rev.pol.trig.se.minus-replay-only.max\endcsname{$-$0.6}
\expandafter\def\csname val@rev.pol.trig.se.minus-replay-only.min\endcsname{-0.025}
\expandafter\def\csname cnt@rev.pol.trig.se.minus-replay-only.min\endcsname{0}
\expandafter\def\csname pct@rev.pol.trig.se.minus-replay-only.min\endcsname{$-$2.5}
\expandafter\def\csname ppt@rev.pol.trig.se.minus-replay-only.min\endcsname{$-$2.5}
\expandafter\def\csname val@rev.pol.trig.se.minus-replay-only.n-ci-above-0\endcsname{0}
\expandafter\def\csname cnt@rev.pol.trig.se.minus-replay-only.n-ci-above-0\endcsname{0}
\expandafter\def\csname pct@rev.pol.trig.se.minus-replay-only.n-ci-above-0\endcsname{0.0}
\expandafter\def\csname ppt@rev.pol.trig.se.minus-replay-only.n-ci-above-0\endcsname{0.0}
\expandafter\def\csname val@rev.pol.trig.se.minus-replay-only.n-ci-below-0\endcsname{3}
\expandafter\def\csname cnt@rev.pol.trig.se.minus-replay-only.n-ci-below-0\endcsname{3}
\expandafter\def\csname pct@rev.pol.trig.se.minus-replay-only.n-ci-below-0\endcsname{300.0}
\expandafter\def\csname ppt@rev.pol.trig.se.minus-replay-only.n-ci-below-0\endcsname{$+$300.0}
\expandafter\def\csname val@rev.pol.trig.se.minus-replay-ppi-uniform.max\endcsname{0.008}
\expandafter\def\csname cnt@rev.pol.trig.se.minus-replay-ppi-uniform.max\endcsname{0}
\expandafter\def\csname pct@rev.pol.trig.se.minus-replay-ppi-uniform.max\endcsname{0.8}
\expandafter\def\csname ppt@rev.pol.trig.se.minus-replay-ppi-uniform.max\endcsname{$+$0.8}
\expandafter\def\csname val@rev.pol.trig.se.minus-replay-ppi-uniform.min\endcsname{0.000}
\expandafter\def\csname cnt@rev.pol.trig.se.minus-replay-ppi-uniform.min\endcsname{0}
\expandafter\def\csname pct@rev.pol.trig.se.minus-replay-ppi-uniform.min\endcsname{0.0}
\expandafter\def\csname ppt@rev.pol.trig.se.minus-replay-ppi-uniform.min\endcsname{0.0}
\expandafter\def\csname val@rev.pol.trig.se.minus-replay-ppi-uniform.n-ci-above-0\endcsname{2}
\expandafter\def\csname cnt@rev.pol.trig.se.minus-replay-ppi-uniform.n-ci-above-0\endcsname{2}
\expandafter\def\csname pct@rev.pol.trig.se.minus-replay-ppi-uniform.n-ci-above-0\endcsname{200.0}
\expandafter\def\csname ppt@rev.pol.trig.se.minus-replay-ppi-uniform.n-ci-above-0\endcsname{$+$200.0}
\expandafter\def\csname val@rev.pol.trig.se.minus-replay-ppi-uniform.n-ci-below-0\endcsname{0}
\expandafter\def\csname cnt@rev.pol.trig.se.minus-replay-ppi-uniform.n-ci-below-0\endcsname{0}
\expandafter\def\csname pct@rev.pol.trig.se.minus-replay-ppi-uniform.n-ci-below-0\endcsname{0.0}
\expandafter\def\csname ppt@rev.pol.trig.se.minus-replay-ppi-uniform.n-ci-below-0\endcsname{0.0}
\expandafter\def\csname val@rev.pol.trig.se.n-doubt-total\endcsname{14}
\expandafter\def\csname cnt@rev.pol.trig.se.n-doubt-total\endcsname{14}
\expandafter\def\csname pct@rev.pol.trig.se.n-doubt-total\endcsname{1400.0}
\expandafter\def\csname ppt@rev.pol.trig.se.n-doubt-total\endcsname{$+$1400.0}
\expandafter\def\csname val@rev.pol.trig.se.sspro.p1.cost\endcsname{128.876}
\expandafter\def\csname cnt@rev.pol.trig.se.sspro.p1.cost\endcsname{129}
\expandafter\def\csname pct@rev.pol.trig.se.sspro.p1.cost\endcsname{12887.6}
\expandafter\def\csname ppt@rev.pol.trig.se.sspro.p1.cost\endcsname{$+$12887.6}
\expandafter\def\csname val@rev.pol.trig.se.sspro.p1.keep\endcsname{0.949}
\expandafter\def\csname cnt@rev.pol.trig.se.sspro.p1.keep\endcsname{1}
\expandafter\def\csname pct@rev.pol.trig.se.sspro.p1.keep\endcsname{94.9}
\expandafter\def\csname ppt@rev.pol.trig.se.sspro.p1.keep\endcsname{$+$94.9}
\expandafter\def\csname val@rev.pol.trig.se.sspro.p1.minus-rcr-stratified.ci95.0\endcsname{-0.007}
\expandafter\def\csname cnt@rev.pol.trig.se.sspro.p1.minus-rcr-stratified.ci95.0\endcsname{0}
\expandafter\def\csname pct@rev.pol.trig.se.sspro.p1.minus-rcr-stratified.ci95.0\endcsname{$-$0.7}
\expandafter\def\csname ppt@rev.pol.trig.se.sspro.p1.minus-rcr-stratified.ci95.0\endcsname{$-$0.7}
\expandafter\def\csname val@rev.pol.trig.se.sspro.p1.minus-rcr-stratified.ci95.1\endcsname{-0.000}
\expandafter\def\csname cnt@rev.pol.trig.se.sspro.p1.minus-rcr-stratified.ci95.1\endcsname{0}
\expandafter\def\csname pct@rev.pol.trig.se.sspro.p1.minus-rcr-stratified.ci95.1\endcsname{0.0}
\expandafter\def\csname ppt@rev.pol.trig.se.sspro.p1.minus-rcr-stratified.ci95.1\endcsname{0.0}
\expandafter\def\csname val@rev.pol.trig.se.sspro.p1.minus-rcr-stratified.mean\endcsname{-0.003}
\expandafter\def\csname cnt@rev.pol.trig.se.sspro.p1.minus-rcr-stratified.mean\endcsname{0}
\expandafter\def\csname pct@rev.pol.trig.se.sspro.p1.minus-rcr-stratified.mean\endcsname{$-$0.3}
\expandafter\def\csname ppt@rev.pol.trig.se.sspro.p1.minus-rcr-stratified.mean\endcsname{$-$0.3}
\expandafter\def\csname val@rev.pol.trig.se.sspro.p1.minus-replay-only.ci95.0\endcsname{-0.014}
\expandafter\def\csname cnt@rev.pol.trig.se.sspro.p1.minus-replay-only.ci95.0\endcsname{0}
\expandafter\def\csname pct@rev.pol.trig.se.sspro.p1.minus-replay-only.ci95.0\endcsname{$-$1.4}
\expandafter\def\csname ppt@rev.pol.trig.se.sspro.p1.minus-replay-only.ci95.0\endcsname{$-$1.4}
\expandafter\def\csname val@rev.pol.trig.se.sspro.p1.minus-replay-only.ci95.1\endcsname{-0.001}
\expandafter\def\csname cnt@rev.pol.trig.se.sspro.p1.minus-replay-only.ci95.1\endcsname{0}
\expandafter\def\csname pct@rev.pol.trig.se.sspro.p1.minus-replay-only.ci95.1\endcsname{$-$0.1}
\expandafter\def\csname ppt@rev.pol.trig.se.sspro.p1.minus-replay-only.ci95.1\endcsname{$-$0.1}
\expandafter\def\csname val@rev.pol.trig.se.sspro.p1.minus-replay-only.mean\endcsname{-0.007}
\expandafter\def\csname cnt@rev.pol.trig.se.sspro.p1.minus-replay-only.mean\endcsname{0}
\expandafter\def\csname pct@rev.pol.trig.se.sspro.p1.minus-replay-only.mean\endcsname{$-$0.7}
\expandafter\def\csname ppt@rev.pol.trig.se.sspro.p1.minus-replay-only.mean\endcsname{$-$0.7}
\expandafter\def\csname val@rev.pol.trig.se.sspro.p1.minus-replay-ppi-uniform.ci95.0\endcsname{0.000}
\expandafter\def\csname cnt@rev.pol.trig.se.sspro.p1.minus-replay-ppi-uniform.ci95.0\endcsname{0}
\expandafter\def\csname pct@rev.pol.trig.se.sspro.p1.minus-replay-ppi-uniform.ci95.0\endcsname{0.0}
\expandafter\def\csname ppt@rev.pol.trig.se.sspro.p1.minus-replay-ppi-uniform.ci95.0\endcsname{0.0}
\expandafter\def\csname val@rev.pol.trig.se.sspro.p1.minus-replay-ppi-uniform.ci95.1\endcsname{0.000}
\expandafter\def\csname cnt@rev.pol.trig.se.sspro.p1.minus-replay-ppi-uniform.ci95.1\endcsname{0}
\expandafter\def\csname pct@rev.pol.trig.se.sspro.p1.minus-replay-ppi-uniform.ci95.1\endcsname{0.0}
\expandafter\def\csname ppt@rev.pol.trig.se.sspro.p1.minus-replay-ppi-uniform.ci95.1\endcsname{0.0}
\expandafter\def\csname val@rev.pol.trig.se.sspro.p1.minus-replay-ppi-uniform.mean\endcsname{0.000}
\expandafter\def\csname cnt@rev.pol.trig.se.sspro.p1.minus-replay-ppi-uniform.mean\endcsname{0}
\expandafter\def\csname pct@rev.pol.trig.se.sspro.p1.minus-replay-ppi-uniform.mean\endcsname{0.0}
\expandafter\def\csname ppt@rev.pol.trig.se.sspro.p1.minus-replay-ppi-uniform.mean\endcsname{0.0}
\expandafter\def\csname val@rev.pol.trig.se.sspro.p1.n-doubt\endcsname{4}
\expandafter\def\csname cnt@rev.pol.trig.se.sspro.p1.n-doubt\endcsname{4}
\expandafter\def\csname pct@rev.pol.trig.se.sspro.p1.n-doubt\endcsname{400.0}
\expandafter\def\csname ppt@rev.pol.trig.se.sspro.p1.n-doubt\endcsname{$+$400.0}
\expandafter\def\csname val@rev.pol.trig.se.sspro.p1.score\endcsname{0.852}
\expandafter\def\csname cnt@rev.pol.trig.se.sspro.p1.score\endcsname{1}
\expandafter\def\csname pct@rev.pol.trig.se.sspro.p1.score\endcsname{85.2}
\expandafter\def\csname ppt@rev.pol.trig.se.sspro.p1.score\endcsname{$+$85.2}
\expandafter\def\csname val@rev.pol.trig.se.sspro.p1.sd\endcsname{0.015}
\expandafter\def\csname cnt@rev.pol.trig.se.sspro.p1.sd\endcsname{0}
\expandafter\def\csname pct@rev.pol.trig.se.sspro.p1.sd\endcsname{1.5}
\expandafter\def\csname ppt@rev.pol.trig.se.sspro.p1.sd\endcsname{$+$1.5}
\expandafter\def\csname val@rev.pol.trig.se.sspro.p2.cost\endcsname{131.666}
\expandafter\def\csname cnt@rev.pol.trig.se.sspro.p2.cost\endcsname{132}
\expandafter\def\csname pct@rev.pol.trig.se.sspro.p2.cost\endcsname{13166.6}
\expandafter\def\csname ppt@rev.pol.trig.se.sspro.p2.cost\endcsname{$+$13166.6}
\expandafter\def\csname val@rev.pol.trig.se.sspro.p2.keep\endcsname{0.924}
\expandafter\def\csname cnt@rev.pol.trig.se.sspro.p2.keep\endcsname{1}
\expandafter\def\csname pct@rev.pol.trig.se.sspro.p2.keep\endcsname{92.4}
\expandafter\def\csname ppt@rev.pol.trig.se.sspro.p2.keep\endcsname{$+$92.4}
\expandafter\def\csname val@rev.pol.trig.se.sspro.p2.minus-rcr-stratified.ci95.0\endcsname{0.001}
\expandafter\def\csname cnt@rev.pol.trig.se.sspro.p2.minus-rcr-stratified.ci95.0\endcsname{0}
\expandafter\def\csname pct@rev.pol.trig.se.sspro.p2.minus-rcr-stratified.ci95.0\endcsname{0.1}
\expandafter\def\csname ppt@rev.pol.trig.se.sspro.p2.minus-rcr-stratified.ci95.0\endcsname{$+$0.1}
\expandafter\def\csname val@rev.pol.trig.se.sspro.p2.minus-rcr-stratified.ci95.1\endcsname{0.007}
\expandafter\def\csname cnt@rev.pol.trig.se.sspro.p2.minus-rcr-stratified.ci95.1\endcsname{0}
\expandafter\def\csname pct@rev.pol.trig.se.sspro.p2.minus-rcr-stratified.ci95.1\endcsname{0.7}
\expandafter\def\csname ppt@rev.pol.trig.se.sspro.p2.minus-rcr-stratified.ci95.1\endcsname{$+$0.7}
\expandafter\def\csname val@rev.pol.trig.se.sspro.p2.minus-rcr-stratified.mean\endcsname{0.004}
\expandafter\def\csname cnt@rev.pol.trig.se.sspro.p2.minus-rcr-stratified.mean\endcsname{0}
\expandafter\def\csname pct@rev.pol.trig.se.sspro.p2.minus-rcr-stratified.mean\endcsname{0.4}
\expandafter\def\csname ppt@rev.pol.trig.se.sspro.p2.minus-rcr-stratified.mean\endcsname{$+$0.4}
\expandafter\def\csname val@rev.pol.trig.se.sspro.p2.minus-replay-only.ci95.0\endcsname{-0.017}
\expandafter\def\csname cnt@rev.pol.trig.se.sspro.p2.minus-replay-only.ci95.0\endcsname{0}
\expandafter\def\csname pct@rev.pol.trig.se.sspro.p2.minus-replay-only.ci95.0\endcsname{$-$1.7}
\expandafter\def\csname ppt@rev.pol.trig.se.sspro.p2.minus-replay-only.ci95.0\endcsname{$-$1.7}
\expandafter\def\csname val@rev.pol.trig.se.sspro.p2.minus-replay-only.ci95.1\endcsname{-0.006}
\expandafter\def\csname cnt@rev.pol.trig.se.sspro.p2.minus-replay-only.ci95.1\endcsname{0}
\expandafter\def\csname pct@rev.pol.trig.se.sspro.p2.minus-replay-only.ci95.1\endcsname{$-$0.6}
\expandafter\def\csname ppt@rev.pol.trig.se.sspro.p2.minus-replay-only.ci95.1\endcsname{$-$0.6}
\expandafter\def\csname val@rev.pol.trig.se.sspro.p2.minus-replay-only.mean\endcsname{-0.012}
\expandafter\def\csname cnt@rev.pol.trig.se.sspro.p2.minus-replay-only.mean\endcsname{0}
\expandafter\def\csname pct@rev.pol.trig.se.sspro.p2.minus-replay-only.mean\endcsname{$-$1.2}
\expandafter\def\csname ppt@rev.pol.trig.se.sspro.p2.minus-replay-only.mean\endcsname{$-$1.2}
\expandafter\def\csname val@rev.pol.trig.se.sspro.p2.minus-replay-ppi-uniform.ci95.0\endcsname{0.002}
\expandafter\def\csname cnt@rev.pol.trig.se.sspro.p2.minus-replay-ppi-uniform.ci95.0\endcsname{0}
\expandafter\def\csname pct@rev.pol.trig.se.sspro.p2.minus-replay-ppi-uniform.ci95.0\endcsname{0.2}
\expandafter\def\csname ppt@rev.pol.trig.se.sspro.p2.minus-replay-ppi-uniform.ci95.0\endcsname{$+$0.2}
\expandafter\def\csname val@rev.pol.trig.se.sspro.p2.minus-replay-ppi-uniform.ci95.1\endcsname{0.006}
\expandafter\def\csname cnt@rev.pol.trig.se.sspro.p2.minus-replay-ppi-uniform.ci95.1\endcsname{0}
\expandafter\def\csname pct@rev.pol.trig.se.sspro.p2.minus-replay-ppi-uniform.ci95.1\endcsname{0.6}
\expandafter\def\csname ppt@rev.pol.trig.se.sspro.p2.minus-replay-ppi-uniform.ci95.1\endcsname{$+$0.6}
\expandafter\def\csname val@rev.pol.trig.se.sspro.p2.minus-replay-ppi-uniform.mean\endcsname{0.004}
\expandafter\def\csname cnt@rev.pol.trig.se.sspro.p2.minus-replay-ppi-uniform.mean\endcsname{0}
\expandafter\def\csname pct@rev.pol.trig.se.sspro.p2.minus-replay-ppi-uniform.mean\endcsname{0.4}
\expandafter\def\csname ppt@rev.pol.trig.se.sspro.p2.minus-replay-ppi-uniform.mean\endcsname{$+$0.4}
\expandafter\def\csname val@rev.pol.trig.se.sspro.p2.n-doubt\endcsname{3}
\expandafter\def\csname cnt@rev.pol.trig.se.sspro.p2.n-doubt\endcsname{3}
\expandafter\def\csname pct@rev.pol.trig.se.sspro.p2.n-doubt\endcsname{300.0}
\expandafter\def\csname ppt@rev.pol.trig.se.sspro.p2.n-doubt\endcsname{$+$300.0}
\expandafter\def\csname val@rev.pol.trig.se.sspro.p2.score\endcsname{0.894}
\expandafter\def\csname cnt@rev.pol.trig.se.sspro.p2.score\endcsname{1}
\expandafter\def\csname pct@rev.pol.trig.se.sspro.p2.score\endcsname{89.4}
\expandafter\def\csname ppt@rev.pol.trig.se.sspro.p2.score\endcsname{$+$89.4}
\expandafter\def\csname val@rev.pol.trig.se.sspro.p2.sd\endcsname{0.018}
\expandafter\def\csname cnt@rev.pol.trig.se.sspro.p2.sd\endcsname{0}
\expandafter\def\csname pct@rev.pol.trig.se.sspro.p2.sd\endcsname{1.8}
\expandafter\def\csname ppt@rev.pol.trig.se.sspro.p2.sd\endcsname{$+$1.8}
\expandafter\def\csname val@rev.pol.trig.se.uiv.p1.cost\endcsname{248.251}
\expandafter\def\csname cnt@rev.pol.trig.se.uiv.p1.cost\endcsname{248}
\expandafter\def\csname pct@rev.pol.trig.se.uiv.p1.cost\endcsname{24825.1}
\expandafter\def\csname ppt@rev.pol.trig.se.uiv.p1.cost\endcsname{$+$24825.1}
\expandafter\def\csname val@rev.pol.trig.se.uiv.p1.keep\endcsname{0.347}
\expandafter\def\csname cnt@rev.pol.trig.se.uiv.p1.keep\endcsname{0}
\expandafter\def\csname pct@rev.pol.trig.se.uiv.p1.keep\endcsname{34.7}
\expandafter\def\csname ppt@rev.pol.trig.se.uiv.p1.keep\endcsname{$+$34.7}
\expandafter\def\csname val@rev.pol.trig.se.uiv.p1.minus-rcr-stratified.ci95.0\endcsname{0.001}
\expandafter\def\csname cnt@rev.pol.trig.se.uiv.p1.minus-rcr-stratified.ci95.0\endcsname{0}
\expandafter\def\csname pct@rev.pol.trig.se.uiv.p1.minus-rcr-stratified.ci95.0\endcsname{0.1}
\expandafter\def\csname ppt@rev.pol.trig.se.uiv.p1.minus-rcr-stratified.ci95.0\endcsname{$+$0.1}
\expandafter\def\csname val@rev.pol.trig.se.uiv.p1.minus-rcr-stratified.ci95.1\endcsname{0.007}
\expandafter\def\csname cnt@rev.pol.trig.se.uiv.p1.minus-rcr-stratified.ci95.1\endcsname{0}
\expandafter\def\csname pct@rev.pol.trig.se.uiv.p1.minus-rcr-stratified.ci95.1\endcsname{0.7}
\expandafter\def\csname ppt@rev.pol.trig.se.uiv.p1.minus-rcr-stratified.ci95.1\endcsname{$+$0.7}
\expandafter\def\csname val@rev.pol.trig.se.uiv.p1.minus-rcr-stratified.mean\endcsname{0.004}
\expandafter\def\csname cnt@rev.pol.trig.se.uiv.p1.minus-rcr-stratified.mean\endcsname{0}
\expandafter\def\csname pct@rev.pol.trig.se.uiv.p1.minus-rcr-stratified.mean\endcsname{0.4}
\expandafter\def\csname ppt@rev.pol.trig.se.uiv.p1.minus-rcr-stratified.mean\endcsname{$+$0.4}
\expandafter\def\csname val@rev.pol.trig.se.uiv.p1.minus-replay-only.ci95.0\endcsname{-0.039}
\expandafter\def\csname cnt@rev.pol.trig.se.uiv.p1.minus-replay-only.ci95.0\endcsname{0}
\expandafter\def\csname pct@rev.pol.trig.se.uiv.p1.minus-replay-only.ci95.0\endcsname{$-$3.9}
\expandafter\def\csname ppt@rev.pol.trig.se.uiv.p1.minus-replay-only.ci95.0\endcsname{$-$3.9}
\expandafter\def\csname val@rev.pol.trig.se.uiv.p1.minus-replay-only.ci95.1\endcsname{-0.012}
\expandafter\def\csname cnt@rev.pol.trig.se.uiv.p1.minus-replay-only.ci95.1\endcsname{0}
\expandafter\def\csname pct@rev.pol.trig.se.uiv.p1.minus-replay-only.ci95.1\endcsname{$-$1.2}
\expandafter\def\csname ppt@rev.pol.trig.se.uiv.p1.minus-replay-only.ci95.1\endcsname{$-$1.2}
\expandafter\def\csname val@rev.pol.trig.se.uiv.p1.minus-replay-only.mean\endcsname{-0.025}
\expandafter\def\csname cnt@rev.pol.trig.se.uiv.p1.minus-replay-only.mean\endcsname{0}
\expandafter\def\csname pct@rev.pol.trig.se.uiv.p1.minus-replay-only.mean\endcsname{$-$2.5}
\expandafter\def\csname ppt@rev.pol.trig.se.uiv.p1.minus-replay-only.mean\endcsname{$-$2.5}
\expandafter\def\csname val@rev.pol.trig.se.uiv.p1.minus-replay-ppi-uniform.ci95.0\endcsname{0.003}
\expandafter\def\csname cnt@rev.pol.trig.se.uiv.p1.minus-replay-ppi-uniform.ci95.0\endcsname{0}
\expandafter\def\csname pct@rev.pol.trig.se.uiv.p1.minus-replay-ppi-uniform.ci95.0\endcsname{0.3}
\expandafter\def\csname ppt@rev.pol.trig.se.uiv.p1.minus-replay-ppi-uniform.ci95.0\endcsname{$+$0.3}
\expandafter\def\csname val@rev.pol.trig.se.uiv.p1.minus-replay-ppi-uniform.ci95.1\endcsname{0.013}
\expandafter\def\csname cnt@rev.pol.trig.se.uiv.p1.minus-replay-ppi-uniform.ci95.1\endcsname{0}
\expandafter\def\csname pct@rev.pol.trig.se.uiv.p1.minus-replay-ppi-uniform.ci95.1\endcsname{1.3}
\expandafter\def\csname ppt@rev.pol.trig.se.uiv.p1.minus-replay-ppi-uniform.ci95.1\endcsname{$+$1.3}
\expandafter\def\csname val@rev.pol.trig.se.uiv.p1.minus-replay-ppi-uniform.mean\endcsname{0.008}
\expandafter\def\csname cnt@rev.pol.trig.se.uiv.p1.minus-replay-ppi-uniform.mean\endcsname{0}
\expandafter\def\csname pct@rev.pol.trig.se.uiv.p1.minus-replay-ppi-uniform.mean\endcsname{0.8}
\expandafter\def\csname ppt@rev.pol.trig.se.uiv.p1.minus-replay-ppi-uniform.mean\endcsname{$+$0.8}
\expandafter\def\csname val@rev.pol.trig.se.uiv.p1.n-doubt\endcsname{3}
\expandafter\def\csname cnt@rev.pol.trig.se.uiv.p1.n-doubt\endcsname{3}
\expandafter\def\csname pct@rev.pol.trig.se.uiv.p1.n-doubt\endcsname{300.0}
\expandafter\def\csname ppt@rev.pol.trig.se.uiv.p1.n-doubt\endcsname{$+$300.0}
\expandafter\def\csname val@rev.pol.trig.se.uiv.p1.score\endcsname{0.809}
\expandafter\def\csname cnt@rev.pol.trig.se.uiv.p1.score\endcsname{1}
\expandafter\def\csname pct@rev.pol.trig.se.uiv.p1.score\endcsname{80.9}
\expandafter\def\csname ppt@rev.pol.trig.se.uiv.p1.score\endcsname{$+$80.9}
\expandafter\def\csname val@rev.pol.trig.se.uiv.p1.sd\endcsname{0.017}
\expandafter\def\csname cnt@rev.pol.trig.se.uiv.p1.sd\endcsname{0}
\expandafter\def\csname pct@rev.pol.trig.se.uiv.p1.sd\endcsname{1.7}
\expandafter\def\csname ppt@rev.pol.trig.se.uiv.p1.sd\endcsname{$+$1.7}
\expandafter\def\csname val@rev.pol.trig.se.uiv.p2.cost\endcsname{255.503}
\expandafter\def\csname cnt@rev.pol.trig.se.uiv.p2.cost\endcsname{256}
\expandafter\def\csname pct@rev.pol.trig.se.uiv.p2.cost\endcsname{25550.3}
\expandafter\def\csname ppt@rev.pol.trig.se.uiv.p2.cost\endcsname{$+$25550.3}
\expandafter\def\csname val@rev.pol.trig.se.uiv.p2.keep\endcsname{0.580}
\expandafter\def\csname cnt@rev.pol.trig.se.uiv.p2.keep\endcsname{1}
\expandafter\def\csname pct@rev.pol.trig.se.uiv.p2.keep\endcsname{58.0}
\expandafter\def\csname ppt@rev.pol.trig.se.uiv.p2.keep\endcsname{$+$58.0}
\expandafter\def\csname val@rev.pol.trig.se.uiv.p2.minus-rcr-stratified.ci95.0\endcsname{-0.002}
\expandafter\def\csname cnt@rev.pol.trig.se.uiv.p2.minus-rcr-stratified.ci95.0\endcsname{0}
\expandafter\def\csname pct@rev.pol.trig.se.uiv.p2.minus-rcr-stratified.ci95.0\endcsname{$-$0.2}
\expandafter\def\csname ppt@rev.pol.trig.se.uiv.p2.minus-rcr-stratified.ci95.0\endcsname{$-$0.2}
\expandafter\def\csname val@rev.pol.trig.se.uiv.p2.minus-rcr-stratified.ci95.1\endcsname{-0.000}
\expandafter\def\csname cnt@rev.pol.trig.se.uiv.p2.minus-rcr-stratified.ci95.1\endcsname{0}
\expandafter\def\csname pct@rev.pol.trig.se.uiv.p2.minus-rcr-stratified.ci95.1\endcsname{0.0}
\expandafter\def\csname ppt@rev.pol.trig.se.uiv.p2.minus-rcr-stratified.ci95.1\endcsname{0.0}
\expandafter\def\csname val@rev.pol.trig.se.uiv.p2.minus-rcr-stratified.mean\endcsname{-0.001}
\expandafter\def\csname cnt@rev.pol.trig.se.uiv.p2.minus-rcr-stratified.mean\endcsname{0}
\expandafter\def\csname pct@rev.pol.trig.se.uiv.p2.minus-rcr-stratified.mean\endcsname{$-$0.1}
\expandafter\def\csname ppt@rev.pol.trig.se.uiv.p2.minus-rcr-stratified.mean\endcsname{$-$0.1}
\expandafter\def\csname val@rev.pol.trig.se.uiv.p2.minus-replay-only.ci95.0\endcsname{-0.014}
\expandafter\def\csname cnt@rev.pol.trig.se.uiv.p2.minus-replay-only.ci95.0\endcsname{0}
\expandafter\def\csname pct@rev.pol.trig.se.uiv.p2.minus-replay-only.ci95.0\endcsname{$-$1.4}
\expandafter\def\csname ppt@rev.pol.trig.se.uiv.p2.minus-replay-only.ci95.0\endcsname{$-$1.4}
\expandafter\def\csname val@rev.pol.trig.se.uiv.p2.minus-replay-only.ci95.1\endcsname{0.002}
\expandafter\def\csname cnt@rev.pol.trig.se.uiv.p2.minus-replay-only.ci95.1\endcsname{0}
\expandafter\def\csname pct@rev.pol.trig.se.uiv.p2.minus-replay-only.ci95.1\endcsname{0.2}
\expandafter\def\csname ppt@rev.pol.trig.se.uiv.p2.minus-replay-only.ci95.1\endcsname{$+$0.2}
\expandafter\def\csname val@rev.pol.trig.se.uiv.p2.minus-replay-only.mean\endcsname{-0.006}
\expandafter\def\csname cnt@rev.pol.trig.se.uiv.p2.minus-replay-only.mean\endcsname{0}
\expandafter\def\csname pct@rev.pol.trig.se.uiv.p2.minus-replay-only.mean\endcsname{$-$0.6}
\expandafter\def\csname ppt@rev.pol.trig.se.uiv.p2.minus-replay-only.mean\endcsname{$-$0.6}
\expandafter\def\csname val@rev.pol.trig.se.uiv.p2.minus-replay-ppi-uniform.ci95.0\endcsname{0.000}
\expandafter\def\csname cnt@rev.pol.trig.se.uiv.p2.minus-replay-ppi-uniform.ci95.0\endcsname{0}
\expandafter\def\csname pct@rev.pol.trig.se.uiv.p2.minus-replay-ppi-uniform.ci95.0\endcsname{0.0}
\expandafter\def\csname ppt@rev.pol.trig.se.uiv.p2.minus-replay-ppi-uniform.ci95.0\endcsname{0.0}
\expandafter\def\csname val@rev.pol.trig.se.uiv.p2.minus-replay-ppi-uniform.ci95.1\endcsname{0.000}
\expandafter\def\csname cnt@rev.pol.trig.se.uiv.p2.minus-replay-ppi-uniform.ci95.1\endcsname{0}
\expandafter\def\csname pct@rev.pol.trig.se.uiv.p2.minus-replay-ppi-uniform.ci95.1\endcsname{0.0}
\expandafter\def\csname ppt@rev.pol.trig.se.uiv.p2.minus-replay-ppi-uniform.ci95.1\endcsname{0.0}
\expandafter\def\csname val@rev.pol.trig.se.uiv.p2.minus-replay-ppi-uniform.mean\endcsname{0.000}
\expandafter\def\csname cnt@rev.pol.trig.se.uiv.p2.minus-replay-ppi-uniform.mean\endcsname{0}
\expandafter\def\csname pct@rev.pol.trig.se.uiv.p2.minus-replay-ppi-uniform.mean\endcsname{0.0}
\expandafter\def\csname ppt@rev.pol.trig.se.uiv.p2.minus-replay-ppi-uniform.mean\endcsname{0.0}
\expandafter\def\csname val@rev.pol.trig.se.uiv.p2.n-doubt\endcsname{4}
\expandafter\def\csname cnt@rev.pol.trig.se.uiv.p2.n-doubt\endcsname{4}
\expandafter\def\csname pct@rev.pol.trig.se.uiv.p2.n-doubt\endcsname{400.0}
\expandafter\def\csname ppt@rev.pol.trig.se.uiv.p2.n-doubt\endcsname{$+$400.0}
\expandafter\def\csname val@rev.pol.trig.se.uiv.p2.score\endcsname{0.827}
\expandafter\def\csname cnt@rev.pol.trig.se.uiv.p2.score\endcsname{1}
\expandafter\def\csname pct@rev.pol.trig.se.uiv.p2.score\endcsname{82.7}
\expandafter\def\csname ppt@rev.pol.trig.se.uiv.p2.score\endcsname{$+$82.7}
\expandafter\def\csname val@rev.pol.trig.se.uiv.p2.sd\endcsname{0.009}
\expandafter\def\csname cnt@rev.pol.trig.se.uiv.p2.sd\endcsname{0}
\expandafter\def\csname pct@rev.pol.trig.se.uiv.p2.sd\endcsname{0.9}
\expandafter\def\csname ppt@rev.pol.trig.se.uiv.p2.sd\endcsname{$+$0.9}
\expandafter\def\csname val@rev.pol.verify.n-checked\endcsname{6720}
\expandafter\def\csname cnt@rev.pol.verify.n-checked\endcsname{6{,}720}
\expandafter\def\csname pct@rev.pol.verify.n-checked\endcsname{672000.0}
\expandafter\def\csname ppt@rev.pol.verify.n-checked\endcsname{$+$672000.0}
\expandafter\def\csname val@rev.pol.verify.n-mismatches\endcsname{0}
\expandafter\def\csname cnt@rev.pol.verify.n-mismatches\endcsname{0}
\expandafter\def\csname pct@rev.pol.verify.n-mismatches\endcsname{0.0}
\expandafter\def\csname ppt@rev.pol.verify.n-mismatches\endcsname{0.0}
\expandafter\def\csname val@rev.pol.verify.n-settings\endcsname{4}
\expandafter\def\csname cnt@rev.pol.verify.n-settings\endcsname{4}
\expandafter\def\csname pct@rev.pol.verify.n-settings\endcsname{400.0}
\expandafter\def\csname ppt@rev.pol.verify.n-settings\endcsname{$+$400.0}
\expandafter\def\csname val@rev.pol.verify.n-settings-ok\endcsname{4}
\expandafter\def\csname cnt@rev.pol.verify.n-settings-ok\endcsname{4}
\expandafter\def\csname pct@rev.pol.verify.n-settings-ok\endcsname{400.0}
\expandafter\def\csname ppt@rev.pol.verify.n-settings-ok\endcsname{$+$400.0}
\expandafter\def\csname val@rev.pol.verify.sspro.p1.byte-identical\endcsname{1}
\expandafter\def\csname cnt@rev.pol.verify.sspro.p1.byte-identical\endcsname{1}
\expandafter\def\csname pct@rev.pol.verify.sspro.p1.byte-identical\endcsname{100.0}
\expandafter\def\csname ppt@rev.pol.verify.sspro.p1.byte-identical\endcsname{$+$100.0}
\expandafter\def\csname val@rev.pol.verify.sspro.p1.mismatches\endcsname{0}
\expandafter\def\csname cnt@rev.pol.verify.sspro.p1.mismatches\endcsname{0}
\expandafter\def\csname pct@rev.pol.verify.sspro.p1.mismatches\endcsname{0.0}
\expandafter\def\csname ppt@rev.pol.verify.sspro.p1.mismatches\endcsname{0.0}
\expandafter\def\csname val@rev.pol.verify.sspro.p2.byte-identical\endcsname{1}
\expandafter\def\csname cnt@rev.pol.verify.sspro.p2.byte-identical\endcsname{1}
\expandafter\def\csname pct@rev.pol.verify.sspro.p2.byte-identical\endcsname{100.0}
\expandafter\def\csname ppt@rev.pol.verify.sspro.p2.byte-identical\endcsname{$+$100.0}
\expandafter\def\csname val@rev.pol.verify.sspro.p2.mismatches\endcsname{0}
\expandafter\def\csname cnt@rev.pol.verify.sspro.p2.mismatches\endcsname{0}
\expandafter\def\csname pct@rev.pol.verify.sspro.p2.mismatches\endcsname{0.0}
\expandafter\def\csname ppt@rev.pol.verify.sspro.p2.mismatches\endcsname{0.0}
\expandafter\def\csname val@rev.pol.verify.uiv.p1.byte-identical\endcsname{1}
\expandafter\def\csname cnt@rev.pol.verify.uiv.p1.byte-identical\endcsname{1}
\expandafter\def\csname pct@rev.pol.verify.uiv.p1.byte-identical\endcsname{100.0}
\expandafter\def\csname ppt@rev.pol.verify.uiv.p1.byte-identical\endcsname{$+$100.0}
\expandafter\def\csname val@rev.pol.verify.uiv.p1.mismatches\endcsname{0}
\expandafter\def\csname cnt@rev.pol.verify.uiv.p1.mismatches\endcsname{0}
\expandafter\def\csname pct@rev.pol.verify.uiv.p1.mismatches\endcsname{0.0}
\expandafter\def\csname ppt@rev.pol.verify.uiv.p1.mismatches\endcsname{0.0}
\expandafter\def\csname val@rev.pol.verify.uiv.p2.byte-identical\endcsname{1}
\expandafter\def\csname cnt@rev.pol.verify.uiv.p2.byte-identical\endcsname{1}
\expandafter\def\csname pct@rev.pol.verify.uiv.p2.byte-identical\endcsname{100.0}
\expandafter\def\csname ppt@rev.pol.verify.uiv.p2.byte-identical\endcsname{$+$100.0}
\expandafter\def\csname val@rev.pol.verify.uiv.p2.mismatches\endcsname{0}
\expandafter\def\csname cnt@rev.pol.verify.uiv.p2.mismatches\endcsname{0}
\expandafter\def\csname pct@rev.pol.verify.uiv.p2.mismatches\endcsname{0.0}
\expandafter\def\csname ppt@rev.pol.verify.uiv.p2.mismatches\endcsname{0.0}
\expandafter\def\csname val@rev.pol.zr.e7-crosscheck-ok\endcsname{1}
\expandafter\def\csname cnt@rev.pol.zr.e7-crosscheck-ok\endcsname{1}
\expandafter\def\csname pct@rev.pol.zr.e7-crosscheck-ok\endcsname{100.0}
\expandafter\def\csname ppt@rev.pol.zr.e7-crosscheck-ok\endcsname{$+$100.0}
\expandafter\def\csname val@rev.pol.zr.incorrect.best-single-changed\endcsname{0}
\expandafter\def\csname cnt@rev.pol.zr.incorrect.best-single-changed\endcsname{0}
\expandafter\def\csname pct@rev.pol.zr.incorrect.best-single-changed\endcsname{0.0}
\expandafter\def\csname ppt@rev.pol.zr.incorrect.best-single-changed\endcsname{0.0}
\expandafter\def\csname val@rev.pol.zr.incorrect.best-single.primary-delta\endcsname{0.000}
\expandafter\def\csname cnt@rev.pol.zr.incorrect.best-single.primary-delta\endcsname{0}
\expandafter\def\csname pct@rev.pol.zr.incorrect.best-single.primary-delta\endcsname{0.0}
\expandafter\def\csname ppt@rev.pol.zr.incorrect.best-single.primary-delta\endcsname{0.0}
\expandafter\def\csname val@rev.pol.zr.incorrect.full-recert.delta\endcsname{0.000}
\expandafter\def\csname cnt@rev.pol.zr.incorrect.full-recert.delta\endcsname{0}
\expandafter\def\csname pct@rev.pol.zr.incorrect.full-recert.delta\endcsname{0.0}
\expandafter\def\csname ppt@rev.pol.zr.incorrect.full-recert.delta\endcsname{0.0}
\expandafter\def\csname val@rev.pol.zr.incorrect.generic-ppi-source.n-changed\endcsname{8}
\expandafter\def\csname cnt@rev.pol.zr.incorrect.generic-ppi-source.n-changed\endcsname{8}
\expandafter\def\csname pct@rev.pol.zr.incorrect.generic-ppi-source.n-changed\endcsname{800.0}
\expandafter\def\csname ppt@rev.pol.zr.incorrect.generic-ppi-source.n-changed\endcsname{$+$800.0}
\expandafter\def\csname val@rev.pol.zr.incorrect.generic-ppi-source.n-changed-cat\endcsname{8}
\expandafter\def\csname cnt@rev.pol.zr.incorrect.generic-ppi-source.n-changed-cat\endcsname{8}
\expandafter\def\csname pct@rev.pol.zr.incorrect.generic-ppi-source.n-changed-cat\endcsname{800.0}
\expandafter\def\csname ppt@rev.pol.zr.incorrect.generic-ppi-source.n-changed-cat\endcsname{$+$800.0}
\expandafter\def\csname val@rev.pol.zr.incorrect.generic-ppi-source.primary-delta\endcsname{0.009}
\expandafter\def\csname cnt@rev.pol.zr.incorrect.generic-ppi-source.primary-delta\endcsname{0}
\expandafter\def\csname pct@rev.pol.zr.incorrect.generic-ppi-source.primary-delta\endcsname{0.9}
\expandafter\def\csname ppt@rev.pol.zr.incorrect.generic-ppi-source.primary-delta\endcsname{$+$0.9}
\expandafter\def\csname val@rev.pol.zr.incorrect.max-abs-score-delta\endcsname{0.009}
\expandafter\def\csname cnt@rev.pol.zr.incorrect.max-abs-score-delta\endcsname{0}
\expandafter\def\csname pct@rev.pol.zr.incorrect.max-abs-score-delta\endcsname{0.9}
\expandafter\def\csname ppt@rev.pol.zr.incorrect.max-abs-score-delta\endcsname{$+$0.9}
\expandafter\def\csname val@rev.pol.zr.incorrect.n-changed\endcsname{26}
\expandafter\def\csname cnt@rev.pol.zr.incorrect.n-changed\endcsname{26}
\expandafter\def\csname pct@rev.pol.zr.incorrect.n-changed\endcsname{2600.0}
\expandafter\def\csname ppt@rev.pol.zr.incorrect.n-changed\endcsname{$+$2600.0}
\expandafter\def\csname val@rev.pol.zr.incorrect.n-changed-cat\endcsname{26}
\expandafter\def\csname cnt@rev.pol.zr.incorrect.n-changed-cat\endcsname{26}
\expandafter\def\csname pct@rev.pol.zr.incorrect.n-changed-cat\endcsname{2600.0}
\expandafter\def\csname ppt@rev.pol.zr.incorrect.n-changed-cat\endcsname{$+$2600.0}
\expandafter\def\csname val@rev.pol.zr.incorrect.n-total\endcsname{1600}
\expandafter\def\csname cnt@rev.pol.zr.incorrect.n-total\endcsname{1{,}600}
\expandafter\def\csname pct@rev.pol.zr.incorrect.n-total\endcsname{160000.0}
\expandafter\def\csname ppt@rev.pol.zr.incorrect.n-total\endcsname{$+$160000.0}
\expandafter\def\csname val@rev.pol.zr.incorrect.n-total-cat\endcsname{1280}
\expandafter\def\csname cnt@rev.pol.zr.incorrect.n-total-cat\endcsname{1{,}280}
\expandafter\def\csname pct@rev.pol.zr.incorrect.n-total-cat\endcsname{128000.0}
\expandafter\def\csname ppt@rev.pol.zr.incorrect.n-total-cat\endcsname{$+$128000.0}
\expandafter\def\csname val@rev.pol.zr.incorrect.no-budget-changed\endcsname{0}
\expandafter\def\csname cnt@rev.pol.zr.incorrect.no-budget-changed\endcsname{0}
\expandafter\def\csname pct@rev.pol.zr.incorrect.no-budget-changed\endcsname{0.0}
\expandafter\def\csname ppt@rev.pol.zr.incorrect.no-budget-changed\endcsname{0.0}
\expandafter\def\csname val@rev.pol.zr.incorrect.no-budget-changed-cat\endcsname{0}
\expandafter\def\csname cnt@rev.pol.zr.incorrect.no-budget-changed-cat\endcsname{0}
\expandafter\def\csname pct@rev.pol.zr.incorrect.no-budget-changed-cat\endcsname{0.0}
\expandafter\def\csname ppt@rev.pol.zr.incorrect.no-budget-changed-cat\endcsname{0.0}
\expandafter\def\csname val@rev.pol.zr.incorrect.old-decision.delta\endcsname{0.000}
\expandafter\def\csname cnt@rev.pol.zr.incorrect.old-decision.delta\endcsname{0}
\expandafter\def\csname pct@rev.pol.zr.incorrect.old-decision.delta\endcsname{0.0}
\expandafter\def\csname ppt@rev.pol.zr.incorrect.old-decision.delta\endcsname{0.0}
\expandafter\def\csname val@rev.pol.zr.incorrect.rcr-minus-generic-ppi-source.base\endcsname{0.013}
\expandafter\def\csname cnt@rev.pol.zr.incorrect.rcr-minus-generic-ppi-source.base\endcsname{0}
\expandafter\def\csname pct@rev.pol.zr.incorrect.rcr-minus-generic-ppi-source.base\endcsname{1.3}
\expandafter\def\csname ppt@rev.pol.zr.incorrect.rcr-minus-generic-ppi-source.base\endcsname{$+$1.3}
\expandafter\def\csname val@rev.pol.zr.incorrect.rcr-minus-generic-ppi-source.variant\endcsname{0.010}
\expandafter\def\csname cnt@rev.pol.zr.incorrect.rcr-minus-generic-ppi-source.variant\endcsname{0}
\expandafter\def\csname pct@rev.pol.zr.incorrect.rcr-minus-generic-ppi-source.variant\endcsname{1.0}
\expandafter\def\csname ppt@rev.pol.zr.incorrect.rcr-minus-generic-ppi-source.variant\endcsname{$+$1.0}
\expandafter\def\csname val@rev.pol.zr.incorrect.rcr-minus-replay-only.base\endcsname{-0.030}
\expandafter\def\csname cnt@rev.pol.zr.incorrect.rcr-minus-replay-only.base\endcsname{0}
\expandafter\def\csname pct@rev.pol.zr.incorrect.rcr-minus-replay-only.base\endcsname{$-$3.0}
\expandafter\def\csname ppt@rev.pol.zr.incorrect.rcr-minus-replay-only.base\endcsname{$-$3.0}
\expandafter\def\csname val@rev.pol.zr.incorrect.rcr-minus-replay-only.variant\endcsname{-0.025}
\expandafter\def\csname cnt@rev.pol.zr.incorrect.rcr-minus-replay-only.variant\endcsname{0}
\expandafter\def\csname pct@rev.pol.zr.incorrect.rcr-minus-replay-only.variant\endcsname{$-$2.5}
\expandafter\def\csname ppt@rev.pol.zr.incorrect.rcr-minus-replay-only.variant\endcsname{$-$2.5}
\expandafter\def\csname val@rev.pol.zr.incorrect.rcr-minus-replay-ppi-uniform.base\endcsname{0.001}
\expandafter\def\csname cnt@rev.pol.zr.incorrect.rcr-minus-replay-ppi-uniform.base\endcsname{0}
\expandafter\def\csname pct@rev.pol.zr.incorrect.rcr-minus-replay-ppi-uniform.base\endcsname{0.1}
\expandafter\def\csname ppt@rev.pol.zr.incorrect.rcr-minus-replay-ppi-uniform.base\endcsname{$+$0.1}
\expandafter\def\csname val@rev.pol.zr.incorrect.rcr-minus-replay-ppi-uniform.variant\endcsname{0.007}
\expandafter\def\csname cnt@rev.pol.zr.incorrect.rcr-minus-replay-ppi-uniform.variant\endcsname{0}
\expandafter\def\csname pct@rev.pol.zr.incorrect.rcr-minus-replay-ppi-uniform.variant\endcsname{0.7}
\expandafter\def\csname ppt@rev.pol.zr.incorrect.rcr-minus-replay-ppi-uniform.variant\endcsname{$+$0.7}
\expandafter\def\csname val@rev.pol.zr.incorrect.rcr-minus-uniform-fresh.base\endcsname{0.007}
\expandafter\def\csname cnt@rev.pol.zr.incorrect.rcr-minus-uniform-fresh.base\endcsname{0}
\expandafter\def\csname pct@rev.pol.zr.incorrect.rcr-minus-uniform-fresh.base\endcsname{0.7}
\expandafter\def\csname ppt@rev.pol.zr.incorrect.rcr-minus-uniform-fresh.base\endcsname{$+$0.7}
\expandafter\def\csname val@rev.pol.zr.incorrect.rcr-minus-uniform-fresh.variant\endcsname{0.012}
\expandafter\def\csname cnt@rev.pol.zr.incorrect.rcr-minus-uniform-fresh.variant\endcsname{0}
\expandafter\def\csname pct@rev.pol.zr.incorrect.rcr-minus-uniform-fresh.variant\endcsname{1.2}
\expandafter\def\csname ppt@rev.pol.zr.incorrect.rcr-minus-uniform-fresh.variant\endcsname{$+$1.2}
\expandafter\def\csname val@rev.pol.zr.incorrect.rcr-stratified.n-changed\endcsname{18}
\expandafter\def\csname cnt@rev.pol.zr.incorrect.rcr-stratified.n-changed\endcsname{18}
\expandafter\def\csname pct@rev.pol.zr.incorrect.rcr-stratified.n-changed\endcsname{1800.0}
\expandafter\def\csname ppt@rev.pol.zr.incorrect.rcr-stratified.n-changed\endcsname{$+$1800.0}
\expandafter\def\csname val@rev.pol.zr.incorrect.rcr-stratified.n-changed-cat\endcsname{18}
\expandafter\def\csname cnt@rev.pol.zr.incorrect.rcr-stratified.n-changed-cat\endcsname{18}
\expandafter\def\csname pct@rev.pol.zr.incorrect.rcr-stratified.n-changed-cat\endcsname{1800.0}
\expandafter\def\csname ppt@rev.pol.zr.incorrect.rcr-stratified.n-changed-cat\endcsname{$+$1800.0}
\expandafter\def\csname val@rev.pol.zr.incorrect.rcr-stratified.primary-delta\endcsname{0.005}
\expandafter\def\csname cnt@rev.pol.zr.incorrect.rcr-stratified.primary-delta\endcsname{0}
\expandafter\def\csname pct@rev.pol.zr.incorrect.rcr-stratified.primary-delta\endcsname{0.5}
\expandafter\def\csname ppt@rev.pol.zr.incorrect.rcr-stratified.primary-delta\endcsname{$+$0.5}
\expandafter\def\csname val@rev.pol.zr.incorrect.replay-only.delta\endcsname{0.000}
\expandafter\def\csname cnt@rev.pol.zr.incorrect.replay-only.delta\endcsname{0}
\expandafter\def\csname pct@rev.pol.zr.incorrect.replay-only.delta\endcsname{0.0}
\expandafter\def\csname ppt@rev.pol.zr.incorrect.replay-only.delta\endcsname{0.0}
\expandafter\def\csname val@rev.pol.zr.incorrect.replay-ppi-uniform.n-changed\endcsname{0}
\expandafter\def\csname cnt@rev.pol.zr.incorrect.replay-ppi-uniform.n-changed\endcsname{0}
\expandafter\def\csname pct@rev.pol.zr.incorrect.replay-ppi-uniform.n-changed\endcsname{0.0}
\expandafter\def\csname ppt@rev.pol.zr.incorrect.replay-ppi-uniform.n-changed\endcsname{0.0}
\expandafter\def\csname val@rev.pol.zr.incorrect.replay-ppi-uniform.n-changed-cat\endcsname{0}
\expandafter\def\csname cnt@rev.pol.zr.incorrect.replay-ppi-uniform.n-changed-cat\endcsname{0}
\expandafter\def\csname pct@rev.pol.zr.incorrect.replay-ppi-uniform.n-changed-cat\endcsname{0.0}
\expandafter\def\csname ppt@rev.pol.zr.incorrect.replay-ppi-uniform.n-changed-cat\endcsname{0.0}
\expandafter\def\csname val@rev.pol.zr.incorrect.replay-ppi-uniform.primary-delta\endcsname{0.000}
\expandafter\def\csname cnt@rev.pol.zr.incorrect.replay-ppi-uniform.primary-delta\endcsname{0}
\expandafter\def\csname pct@rev.pol.zr.incorrect.replay-ppi-uniform.primary-delta\endcsname{0.0}
\expandafter\def\csname ppt@rev.pol.zr.incorrect.replay-ppi-uniform.primary-delta\endcsname{0.0}
\expandafter\def\csname val@rev.pol.zr.incorrect.uniform-fresh.n-changed\endcsname{0}
\expandafter\def\csname cnt@rev.pol.zr.incorrect.uniform-fresh.n-changed\endcsname{0}
\expandafter\def\csname pct@rev.pol.zr.incorrect.uniform-fresh.n-changed\endcsname{0.0}
\expandafter\def\csname ppt@rev.pol.zr.incorrect.uniform-fresh.n-changed\endcsname{0.0}
\expandafter\def\csname val@rev.pol.zr.incorrect.uniform-fresh.n-changed-cat\endcsname{0}
\expandafter\def\csname cnt@rev.pol.zr.incorrect.uniform-fresh.n-changed-cat\endcsname{0}
\expandafter\def\csname pct@rev.pol.zr.incorrect.uniform-fresh.n-changed-cat\endcsname{0.0}
\expandafter\def\csname ppt@rev.pol.zr.incorrect.uniform-fresh.n-changed-cat\endcsname{0.0}
\expandafter\def\csname val@rev.pol.zr.incorrect.uniform-fresh.primary-delta\endcsname{0.000}
\expandafter\def\csname cnt@rev.pol.zr.incorrect.uniform-fresh.primary-delta\endcsname{0}
\expandafter\def\csname pct@rev.pol.zr.incorrect.uniform-fresh.primary-delta\endcsname{0.0}
\expandafter\def\csname ppt@rev.pol.zr.incorrect.uniform-fresh.primary-delta\endcsname{0.0}
\expandafter\def\csname val@rev.pol.zr.missing.best-single-changed\endcsname{0}
\expandafter\def\csname cnt@rev.pol.zr.missing.best-single-changed\endcsname{0}
\expandafter\def\csname pct@rev.pol.zr.missing.best-single-changed\endcsname{0.0}
\expandafter\def\csname ppt@rev.pol.zr.missing.best-single-changed\endcsname{0.0}
\expandafter\def\csname val@rev.pol.zr.missing.best-single.primary-delta\endcsname{0.000}
\expandafter\def\csname cnt@rev.pol.zr.missing.best-single.primary-delta\endcsname{0}
\expandafter\def\csname pct@rev.pol.zr.missing.best-single.primary-delta\endcsname{0.0}
\expandafter\def\csname ppt@rev.pol.zr.missing.best-single.primary-delta\endcsname{0.0}
\expandafter\def\csname val@rev.pol.zr.missing.full-recert.delta\endcsname{0.000}
\expandafter\def\csname cnt@rev.pol.zr.missing.full-recert.delta\endcsname{0}
\expandafter\def\csname pct@rev.pol.zr.missing.full-recert.delta\endcsname{0.0}
\expandafter\def\csname ppt@rev.pol.zr.missing.full-recert.delta\endcsname{0.0}
\expandafter\def\csname val@rev.pol.zr.missing.generic-ppi-source.n-changed\endcsname{87}
\expandafter\def\csname cnt@rev.pol.zr.missing.generic-ppi-source.n-changed\endcsname{87}
\expandafter\def\csname pct@rev.pol.zr.missing.generic-ppi-source.n-changed\endcsname{8700.0}
\expandafter\def\csname ppt@rev.pol.zr.missing.generic-ppi-source.n-changed\endcsname{$+$8700.0}
\expandafter\def\csname val@rev.pol.zr.missing.generic-ppi-source.n-changed-cat\endcsname{62}
\expandafter\def\csname cnt@rev.pol.zr.missing.generic-ppi-source.n-changed-cat\endcsname{62}
\expandafter\def\csname pct@rev.pol.zr.missing.generic-ppi-source.n-changed-cat\endcsname{6200.0}
\expandafter\def\csname ppt@rev.pol.zr.missing.generic-ppi-source.n-changed-cat\endcsname{$+$6200.0}
\expandafter\def\csname val@rev.pol.zr.missing.generic-ppi-source.primary-delta\endcsname{0.007}
\expandafter\def\csname cnt@rev.pol.zr.missing.generic-ppi-source.primary-delta\endcsname{0}
\expandafter\def\csname pct@rev.pol.zr.missing.generic-ppi-source.primary-delta\endcsname{0.7}
\expandafter\def\csname ppt@rev.pol.zr.missing.generic-ppi-source.primary-delta\endcsname{$+$0.7}
\expandafter\def\csname val@rev.pol.zr.missing.max-abs-score-delta\endcsname{0.010}
\expandafter\def\csname cnt@rev.pol.zr.missing.max-abs-score-delta\endcsname{0}
\expandafter\def\csname pct@rev.pol.zr.missing.max-abs-score-delta\endcsname{1.0}
\expandafter\def\csname ppt@rev.pol.zr.missing.max-abs-score-delta\endcsname{$+$1.0}
\expandafter\def\csname val@rev.pol.zr.missing.n-changed\endcsname{275}
\expandafter\def\csname cnt@rev.pol.zr.missing.n-changed\endcsname{275}
\expandafter\def\csname pct@rev.pol.zr.missing.n-changed\endcsname{27500.0}
\expandafter\def\csname ppt@rev.pol.zr.missing.n-changed\endcsname{$+$27500.0}
\expandafter\def\csname val@rev.pol.zr.missing.n-changed-cat\endcsname{196}
\expandafter\def\csname cnt@rev.pol.zr.missing.n-changed-cat\endcsname{196}
\expandafter\def\csname pct@rev.pol.zr.missing.n-changed-cat\endcsname{19600.0}
\expandafter\def\csname ppt@rev.pol.zr.missing.n-changed-cat\endcsname{$+$19600.0}
\expandafter\def\csname val@rev.pol.zr.missing.n-total\endcsname{1600}
\expandafter\def\csname cnt@rev.pol.zr.missing.n-total\endcsname{1{,}600}
\expandafter\def\csname pct@rev.pol.zr.missing.n-total\endcsname{160000.0}
\expandafter\def\csname ppt@rev.pol.zr.missing.n-total\endcsname{$+$160000.0}
\expandafter\def\csname val@rev.pol.zr.missing.n-total-cat\endcsname{1280}
\expandafter\def\csname cnt@rev.pol.zr.missing.n-total-cat\endcsname{1{,}280}
\expandafter\def\csname pct@rev.pol.zr.missing.n-total-cat\endcsname{128000.0}
\expandafter\def\csname ppt@rev.pol.zr.missing.n-total-cat\endcsname{$+$128000.0}
\expandafter\def\csname val@rev.pol.zr.missing.no-budget-changed\endcsname{0}
\expandafter\def\csname cnt@rev.pol.zr.missing.no-budget-changed\endcsname{0}
\expandafter\def\csname pct@rev.pol.zr.missing.no-budget-changed\endcsname{0.0}
\expandafter\def\csname ppt@rev.pol.zr.missing.no-budget-changed\endcsname{0.0}
\expandafter\def\csname val@rev.pol.zr.missing.no-budget-changed-cat\endcsname{0}
\expandafter\def\csname cnt@rev.pol.zr.missing.no-budget-changed-cat\endcsname{0}
\expandafter\def\csname pct@rev.pol.zr.missing.no-budget-changed-cat\endcsname{0.0}
\expandafter\def\csname ppt@rev.pol.zr.missing.no-budget-changed-cat\endcsname{0.0}
\expandafter\def\csname val@rev.pol.zr.missing.old-decision.delta\endcsname{0.000}
\expandafter\def\csname cnt@rev.pol.zr.missing.old-decision.delta\endcsname{0}
\expandafter\def\csname pct@rev.pol.zr.missing.old-decision.delta\endcsname{0.0}
\expandafter\def\csname ppt@rev.pol.zr.missing.old-decision.delta\endcsname{0.0}
\expandafter\def\csname val@rev.pol.zr.missing.rcr-minus-generic-ppi-source.base\endcsname{0.013}
\expandafter\def\csname cnt@rev.pol.zr.missing.rcr-minus-generic-ppi-source.base\endcsname{0}
\expandafter\def\csname pct@rev.pol.zr.missing.rcr-minus-generic-ppi-source.base\endcsname{1.3}
\expandafter\def\csname ppt@rev.pol.zr.missing.rcr-minus-generic-ppi-source.base\endcsname{$+$1.3}
\expandafter\def\csname val@rev.pol.zr.missing.rcr-minus-generic-ppi-source.variant\endcsname{0.008}
\expandafter\def\csname cnt@rev.pol.zr.missing.rcr-minus-generic-ppi-source.variant\endcsname{0}
\expandafter\def\csname pct@rev.pol.zr.missing.rcr-minus-generic-ppi-source.variant\endcsname{0.8}
\expandafter\def\csname ppt@rev.pol.zr.missing.rcr-minus-generic-ppi-source.variant\endcsname{$+$0.8}
\expandafter\def\csname val@rev.pol.zr.missing.rcr-minus-replay-only.base\endcsname{-0.030}
\expandafter\def\csname cnt@rev.pol.zr.missing.rcr-minus-replay-only.base\endcsname{0}
\expandafter\def\csname pct@rev.pol.zr.missing.rcr-minus-replay-only.base\endcsname{$-$3.0}
\expandafter\def\csname ppt@rev.pol.zr.missing.rcr-minus-replay-only.base\endcsname{$-$3.0}
\expandafter\def\csname val@rev.pol.zr.missing.rcr-minus-replay-only.variant\endcsname{-0.028}
\expandafter\def\csname cnt@rev.pol.zr.missing.rcr-minus-replay-only.variant\endcsname{0}
\expandafter\def\csname pct@rev.pol.zr.missing.rcr-minus-replay-only.variant\endcsname{$-$2.8}
\expandafter\def\csname ppt@rev.pol.zr.missing.rcr-minus-replay-only.variant\endcsname{$-$2.8}
\expandafter\def\csname val@rev.pol.zr.missing.rcr-minus-replay-ppi-uniform.base\endcsname{0.001}
\expandafter\def\csname cnt@rev.pol.zr.missing.rcr-minus-replay-ppi-uniform.base\endcsname{0}
\expandafter\def\csname pct@rev.pol.zr.missing.rcr-minus-replay-ppi-uniform.base\endcsname{0.1}
\expandafter\def\csname ppt@rev.pol.zr.missing.rcr-minus-replay-ppi-uniform.base\endcsname{$+$0.1}
\expandafter\def\csname val@rev.pol.zr.missing.rcr-minus-replay-ppi-uniform.variant\endcsname{-0.001}
\expandafter\def\csname cnt@rev.pol.zr.missing.rcr-minus-replay-ppi-uniform.variant\endcsname{0}
\expandafter\def\csname pct@rev.pol.zr.missing.rcr-minus-replay-ppi-uniform.variant\endcsname{$-$0.1}
\expandafter\def\csname ppt@rev.pol.zr.missing.rcr-minus-replay-ppi-uniform.variant\endcsname{$-$0.1}
\expandafter\def\csname val@rev.pol.zr.missing.rcr-minus-uniform-fresh.base\endcsname{0.007}
\expandafter\def\csname cnt@rev.pol.zr.missing.rcr-minus-uniform-fresh.base\endcsname{0}
\expandafter\def\csname pct@rev.pol.zr.missing.rcr-minus-uniform-fresh.base\endcsname{0.7}
\expandafter\def\csname ppt@rev.pol.zr.missing.rcr-minus-uniform-fresh.base\endcsname{$+$0.7}
\expandafter\def\csname val@rev.pol.zr.missing.rcr-minus-uniform-fresh.variant\endcsname{0.007}
\expandafter\def\csname cnt@rev.pol.zr.missing.rcr-minus-uniform-fresh.variant\endcsname{0}
\expandafter\def\csname pct@rev.pol.zr.missing.rcr-minus-uniform-fresh.variant\endcsname{0.7}
\expandafter\def\csname ppt@rev.pol.zr.missing.rcr-minus-uniform-fresh.variant\endcsname{$+$0.7}
\expandafter\def\csname val@rev.pol.zr.missing.rcr-stratified.n-changed\endcsname{40}
\expandafter\def\csname cnt@rev.pol.zr.missing.rcr-stratified.n-changed\endcsname{40}
\expandafter\def\csname pct@rev.pol.zr.missing.rcr-stratified.n-changed\endcsname{4000.0}
\expandafter\def\csname ppt@rev.pol.zr.missing.rcr-stratified.n-changed\endcsname{$+$4000.0}
\expandafter\def\csname val@rev.pol.zr.missing.rcr-stratified.n-changed-cat\endcsname{34}
\expandafter\def\csname cnt@rev.pol.zr.missing.rcr-stratified.n-changed-cat\endcsname{34}
\expandafter\def\csname pct@rev.pol.zr.missing.rcr-stratified.n-changed-cat\endcsname{3400.0}
\expandafter\def\csname ppt@rev.pol.zr.missing.rcr-stratified.n-changed-cat\endcsname{$+$3400.0}
\expandafter\def\csname val@rev.pol.zr.missing.rcr-stratified.primary-delta\endcsname{0.002}
\expandafter\def\csname cnt@rev.pol.zr.missing.rcr-stratified.primary-delta\endcsname{0}
\expandafter\def\csname pct@rev.pol.zr.missing.rcr-stratified.primary-delta\endcsname{0.2}
\expandafter\def\csname ppt@rev.pol.zr.missing.rcr-stratified.primary-delta\endcsname{$+$0.2}
\expandafter\def\csname val@rev.pol.zr.missing.replay-only.delta\endcsname{0.000}
\expandafter\def\csname cnt@rev.pol.zr.missing.replay-only.delta\endcsname{0}
\expandafter\def\csname pct@rev.pol.zr.missing.replay-only.delta\endcsname{0.0}
\expandafter\def\csname ppt@rev.pol.zr.missing.replay-only.delta\endcsname{0.0}
\expandafter\def\csname val@rev.pol.zr.missing.replay-ppi-uniform.n-changed\endcsname{66}
\expandafter\def\csname cnt@rev.pol.zr.missing.replay-ppi-uniform.n-changed\endcsname{66}
\expandafter\def\csname pct@rev.pol.zr.missing.replay-ppi-uniform.n-changed\endcsname{6600.0}
\expandafter\def\csname ppt@rev.pol.zr.missing.replay-ppi-uniform.n-changed\endcsname{$+$6600.0}
\expandafter\def\csname val@rev.pol.zr.missing.replay-ppi-uniform.n-changed-cat\endcsname{46}
\expandafter\def\csname cnt@rev.pol.zr.missing.replay-ppi-uniform.n-changed-cat\endcsname{46}
\expandafter\def\csname pct@rev.pol.zr.missing.replay-ppi-uniform.n-changed-cat\endcsname{4600.0}
\expandafter\def\csname ppt@rev.pol.zr.missing.replay-ppi-uniform.n-changed-cat\endcsname{$+$4600.0}
\expandafter\def\csname val@rev.pol.zr.missing.replay-ppi-uniform.primary-delta\endcsname{0.004}
\expandafter\def\csname cnt@rev.pol.zr.missing.replay-ppi-uniform.primary-delta\endcsname{0}
\expandafter\def\csname pct@rev.pol.zr.missing.replay-ppi-uniform.primary-delta\endcsname{0.4}
\expandafter\def\csname ppt@rev.pol.zr.missing.replay-ppi-uniform.primary-delta\endcsname{$+$0.4}
\expandafter\def\csname val@rev.pol.zr.missing.uniform-fresh.n-changed\endcsname{82}
\expandafter\def\csname cnt@rev.pol.zr.missing.uniform-fresh.n-changed\endcsname{82}
\expandafter\def\csname pct@rev.pol.zr.missing.uniform-fresh.n-changed\endcsname{8200.0}
\expandafter\def\csname ppt@rev.pol.zr.missing.uniform-fresh.n-changed\endcsname{$+$8200.0}
\expandafter\def\csname val@rev.pol.zr.missing.uniform-fresh.n-changed-cat\endcsname{54}
\expandafter\def\csname cnt@rev.pol.zr.missing.uniform-fresh.n-changed-cat\endcsname{54}
\expandafter\def\csname pct@rev.pol.zr.missing.uniform-fresh.n-changed-cat\endcsname{5400.0}
\expandafter\def\csname ppt@rev.pol.zr.missing.uniform-fresh.n-changed-cat\endcsname{$+$5400.0}
\expandafter\def\csname val@rev.pol.zr.missing.uniform-fresh.primary-delta\endcsname{0.001}
\expandafter\def\csname cnt@rev.pol.zr.missing.uniform-fresh.primary-delta\endcsname{0}
\expandafter\def\csname pct@rev.pol.zr.missing.uniform-fresh.primary-delta\endcsname{0.1}
\expandafter\def\csname ppt@rev.pol.zr.missing.uniform-fresh.primary-delta\endcsname{$+$0.1}
\expandafter\def\csname val@rev.pol.zr.missing-sigma.best-single-changed\endcsname{0}
\expandafter\def\csname cnt@rev.pol.zr.missing-sigma.best-single-changed\endcsname{0}
\expandafter\def\csname pct@rev.pol.zr.missing-sigma.best-single-changed\endcsname{0.0}
\expandafter\def\csname ppt@rev.pol.zr.missing-sigma.best-single-changed\endcsname{0.0}
\expandafter\def\csname val@rev.pol.zr.missing-sigma.best-single.primary-delta\endcsname{0.000}
\expandafter\def\csname cnt@rev.pol.zr.missing-sigma.best-single.primary-delta\endcsname{0}
\expandafter\def\csname pct@rev.pol.zr.missing-sigma.best-single.primary-delta\endcsname{0.0}
\expandafter\def\csname ppt@rev.pol.zr.missing-sigma.best-single.primary-delta\endcsname{0.0}
\expandafter\def\csname val@rev.pol.zr.missing-sigma.full-recert.delta\endcsname{0.000}
\expandafter\def\csname cnt@rev.pol.zr.missing-sigma.full-recert.delta\endcsname{0}
\expandafter\def\csname pct@rev.pol.zr.missing-sigma.full-recert.delta\endcsname{0.0}
\expandafter\def\csname ppt@rev.pol.zr.missing-sigma.full-recert.delta\endcsname{0.0}
\expandafter\def\csname val@rev.pol.zr.missing-sigma.generic-ppi-source.n-changed\endcsname{87}
\expandafter\def\csname cnt@rev.pol.zr.missing-sigma.generic-ppi-source.n-changed\endcsname{87}
\expandafter\def\csname pct@rev.pol.zr.missing-sigma.generic-ppi-source.n-changed\endcsname{8700.0}
\expandafter\def\csname ppt@rev.pol.zr.missing-sigma.generic-ppi-source.n-changed\endcsname{$+$8700.0}
\expandafter\def\csname val@rev.pol.zr.missing-sigma.generic-ppi-source.n-changed-cat\endcsname{62}
\expandafter\def\csname cnt@rev.pol.zr.missing-sigma.generic-ppi-source.n-changed-cat\endcsname{62}
\expandafter\def\csname pct@rev.pol.zr.missing-sigma.generic-ppi-source.n-changed-cat\endcsname{6200.0}
\expandafter\def\csname ppt@rev.pol.zr.missing-sigma.generic-ppi-source.n-changed-cat\endcsname{$+$6200.0}
\expandafter\def\csname val@rev.pol.zr.missing-sigma.generic-ppi-source.primary-delta\endcsname{0.007}
\expandafter\def\csname cnt@rev.pol.zr.missing-sigma.generic-ppi-source.primary-delta\endcsname{0}
\expandafter\def\csname pct@rev.pol.zr.missing-sigma.generic-ppi-source.primary-delta\endcsname{0.7}
\expandafter\def\csname ppt@rev.pol.zr.missing-sigma.generic-ppi-source.primary-delta\endcsname{$+$0.7}
\expandafter\def\csname val@rev.pol.zr.missing-sigma.max-abs-score-delta\endcsname{0.010}
\expandafter\def\csname cnt@rev.pol.zr.missing-sigma.max-abs-score-delta\endcsname{0}
\expandafter\def\csname pct@rev.pol.zr.missing-sigma.max-abs-score-delta\endcsname{1.0}
\expandafter\def\csname ppt@rev.pol.zr.missing-sigma.max-abs-score-delta\endcsname{$+$1.0}
\expandafter\def\csname val@rev.pol.zr.missing-sigma.n-changed\endcsname{276}
\expandafter\def\csname cnt@rev.pol.zr.missing-sigma.n-changed\endcsname{276}
\expandafter\def\csname pct@rev.pol.zr.missing-sigma.n-changed\endcsname{27600.0}
\expandafter\def\csname ppt@rev.pol.zr.missing-sigma.n-changed\endcsname{$+$27600.0}
\expandafter\def\csname val@rev.pol.zr.missing-sigma.n-changed-cat\endcsname{197}
\expandafter\def\csname cnt@rev.pol.zr.missing-sigma.n-changed-cat\endcsname{197}
\expandafter\def\csname pct@rev.pol.zr.missing-sigma.n-changed-cat\endcsname{19700.0}
\expandafter\def\csname ppt@rev.pol.zr.missing-sigma.n-changed-cat\endcsname{$+$19700.0}
\expandafter\def\csname val@rev.pol.zr.missing-sigma.n-total\endcsname{1600}
\expandafter\def\csname cnt@rev.pol.zr.missing-sigma.n-total\endcsname{1{,}600}
\expandafter\def\csname pct@rev.pol.zr.missing-sigma.n-total\endcsname{160000.0}
\expandafter\def\csname ppt@rev.pol.zr.missing-sigma.n-total\endcsname{$+$160000.0}
\expandafter\def\csname val@rev.pol.zr.missing-sigma.n-total-cat\endcsname{1280}
\expandafter\def\csname cnt@rev.pol.zr.missing-sigma.n-total-cat\endcsname{1{,}280}
\expandafter\def\csname pct@rev.pol.zr.missing-sigma.n-total-cat\endcsname{128000.0}
\expandafter\def\csname ppt@rev.pol.zr.missing-sigma.n-total-cat\endcsname{$+$128000.0}
\expandafter\def\csname val@rev.pol.zr.missing-sigma.no-budget-changed\endcsname{0}
\expandafter\def\csname cnt@rev.pol.zr.missing-sigma.no-budget-changed\endcsname{0}
\expandafter\def\csname pct@rev.pol.zr.missing-sigma.no-budget-changed\endcsname{0.0}
\expandafter\def\csname ppt@rev.pol.zr.missing-sigma.no-budget-changed\endcsname{0.0}
\expandafter\def\csname val@rev.pol.zr.missing-sigma.no-budget-changed-cat\endcsname{0}
\expandafter\def\csname cnt@rev.pol.zr.missing-sigma.no-budget-changed-cat\endcsname{0}
\expandafter\def\csname pct@rev.pol.zr.missing-sigma.no-budget-changed-cat\endcsname{0.0}
\expandafter\def\csname ppt@rev.pol.zr.missing-sigma.no-budget-changed-cat\endcsname{0.0}
\expandafter\def\csname val@rev.pol.zr.missing-sigma.old-decision.delta\endcsname{0.000}
\expandafter\def\csname cnt@rev.pol.zr.missing-sigma.old-decision.delta\endcsname{0}
\expandafter\def\csname pct@rev.pol.zr.missing-sigma.old-decision.delta\endcsname{0.0}
\expandafter\def\csname ppt@rev.pol.zr.missing-sigma.old-decision.delta\endcsname{0.0}
\expandafter\def\csname val@rev.pol.zr.missing-sigma.rcr-minus-generic-ppi-source.base\endcsname{0.013}
\expandafter\def\csname cnt@rev.pol.zr.missing-sigma.rcr-minus-generic-ppi-source.base\endcsname{0}
\expandafter\def\csname pct@rev.pol.zr.missing-sigma.rcr-minus-generic-ppi-source.base\endcsname{1.3}
\expandafter\def\csname ppt@rev.pol.zr.missing-sigma.rcr-minus-generic-ppi-source.base\endcsname{$+$1.3}
\expandafter\def\csname val@rev.pol.zr.missing-sigma.rcr-minus-generic-ppi-source.variant\endcsname{0.008}
\expandafter\def\csname cnt@rev.pol.zr.missing-sigma.rcr-minus-generic-ppi-source.variant\endcsname{0}
\expandafter\def\csname pct@rev.pol.zr.missing-sigma.rcr-minus-generic-ppi-source.variant\endcsname{0.8}
\expandafter\def\csname ppt@rev.pol.zr.missing-sigma.rcr-minus-generic-ppi-source.variant\endcsname{$+$0.8}
\expandafter\def\csname val@rev.pol.zr.missing-sigma.rcr-minus-replay-only.base\endcsname{-0.030}
\expandafter\def\csname cnt@rev.pol.zr.missing-sigma.rcr-minus-replay-only.base\endcsname{0}
\expandafter\def\csname pct@rev.pol.zr.missing-sigma.rcr-minus-replay-only.base\endcsname{$-$3.0}
\expandafter\def\csname ppt@rev.pol.zr.missing-sigma.rcr-minus-replay-only.base\endcsname{$-$3.0}
\expandafter\def\csname val@rev.pol.zr.missing-sigma.rcr-minus-replay-only.variant\endcsname{-0.028}
\expandafter\def\csname cnt@rev.pol.zr.missing-sigma.rcr-minus-replay-only.variant\endcsname{0}
\expandafter\def\csname pct@rev.pol.zr.missing-sigma.rcr-minus-replay-only.variant\endcsname{$-$2.8}
\expandafter\def\csname ppt@rev.pol.zr.missing-sigma.rcr-minus-replay-only.variant\endcsname{$-$2.8}
\expandafter\def\csname val@rev.pol.zr.missing-sigma.rcr-minus-replay-ppi-uniform.base\endcsname{0.001}
\expandafter\def\csname cnt@rev.pol.zr.missing-sigma.rcr-minus-replay-ppi-uniform.base\endcsname{0}
\expandafter\def\csname pct@rev.pol.zr.missing-sigma.rcr-minus-replay-ppi-uniform.base\endcsname{0.1}
\expandafter\def\csname ppt@rev.pol.zr.missing-sigma.rcr-minus-replay-ppi-uniform.base\endcsname{$+$0.1}
\expandafter\def\csname val@rev.pol.zr.missing-sigma.rcr-minus-replay-ppi-uniform.variant\endcsname{-0.001}
\expandafter\def\csname cnt@rev.pol.zr.missing-sigma.rcr-minus-replay-ppi-uniform.variant\endcsname{0}
\expandafter\def\csname pct@rev.pol.zr.missing-sigma.rcr-minus-replay-ppi-uniform.variant\endcsname{$-$0.1}
\expandafter\def\csname ppt@rev.pol.zr.missing-sigma.rcr-minus-replay-ppi-uniform.variant\endcsname{$-$0.1}
\expandafter\def\csname val@rev.pol.zr.missing-sigma.rcr-minus-uniform-fresh.base\endcsname{0.007}
\expandafter\def\csname cnt@rev.pol.zr.missing-sigma.rcr-minus-uniform-fresh.base\endcsname{0}
\expandafter\def\csname pct@rev.pol.zr.missing-sigma.rcr-minus-uniform-fresh.base\endcsname{0.7}
\expandafter\def\csname ppt@rev.pol.zr.missing-sigma.rcr-minus-uniform-fresh.base\endcsname{$+$0.7}
\expandafter\def\csname val@rev.pol.zr.missing-sigma.rcr-minus-uniform-fresh.variant\endcsname{0.007}
\expandafter\def\csname cnt@rev.pol.zr.missing-sigma.rcr-minus-uniform-fresh.variant\endcsname{0}
\expandafter\def\csname pct@rev.pol.zr.missing-sigma.rcr-minus-uniform-fresh.variant\endcsname{0.7}
\expandafter\def\csname ppt@rev.pol.zr.missing-sigma.rcr-minus-uniform-fresh.variant\endcsname{$+$0.7}
\expandafter\def\csname val@rev.pol.zr.missing-sigma.rcr-stratified.n-changed\endcsname{41}
\expandafter\def\csname cnt@rev.pol.zr.missing-sigma.rcr-stratified.n-changed\endcsname{41}
\expandafter\def\csname pct@rev.pol.zr.missing-sigma.rcr-stratified.n-changed\endcsname{4100.0}
\expandafter\def\csname ppt@rev.pol.zr.missing-sigma.rcr-stratified.n-changed\endcsname{$+$4100.0}
\expandafter\def\csname val@rev.pol.zr.missing-sigma.rcr-stratified.n-changed-cat\endcsname{35}
\expandafter\def\csname cnt@rev.pol.zr.missing-sigma.rcr-stratified.n-changed-cat\endcsname{35}
\expandafter\def\csname pct@rev.pol.zr.missing-sigma.rcr-stratified.n-changed-cat\endcsname{3500.0}
\expandafter\def\csname ppt@rev.pol.zr.missing-sigma.rcr-stratified.n-changed-cat\endcsname{$+$3500.0}
\expandafter\def\csname val@rev.pol.zr.missing-sigma.rcr-stratified.primary-delta\endcsname{0.002}
\expandafter\def\csname cnt@rev.pol.zr.missing-sigma.rcr-stratified.primary-delta\endcsname{0}
\expandafter\def\csname pct@rev.pol.zr.missing-sigma.rcr-stratified.primary-delta\endcsname{0.2}
\expandafter\def\csname ppt@rev.pol.zr.missing-sigma.rcr-stratified.primary-delta\endcsname{$+$0.2}
\expandafter\def\csname val@rev.pol.zr.missing-sigma.replay-only.delta\endcsname{0.000}
\expandafter\def\csname cnt@rev.pol.zr.missing-sigma.replay-only.delta\endcsname{0}
\expandafter\def\csname pct@rev.pol.zr.missing-sigma.replay-only.delta\endcsname{0.0}
\expandafter\def\csname ppt@rev.pol.zr.missing-sigma.replay-only.delta\endcsname{0.0}
\expandafter\def\csname val@rev.pol.zr.missing-sigma.replay-ppi-uniform.n-changed\endcsname{66}
\expandafter\def\csname cnt@rev.pol.zr.missing-sigma.replay-ppi-uniform.n-changed\endcsname{66}
\expandafter\def\csname pct@rev.pol.zr.missing-sigma.replay-ppi-uniform.n-changed\endcsname{6600.0}
\expandafter\def\csname ppt@rev.pol.zr.missing-sigma.replay-ppi-uniform.n-changed\endcsname{$+$6600.0}
\expandafter\def\csname val@rev.pol.zr.missing-sigma.replay-ppi-uniform.n-changed-cat\endcsname{46}
\expandafter\def\csname cnt@rev.pol.zr.missing-sigma.replay-ppi-uniform.n-changed-cat\endcsname{46}
\expandafter\def\csname pct@rev.pol.zr.missing-sigma.replay-ppi-uniform.n-changed-cat\endcsname{4600.0}
\expandafter\def\csname ppt@rev.pol.zr.missing-sigma.replay-ppi-uniform.n-changed-cat\endcsname{$+$4600.0}
\expandafter\def\csname val@rev.pol.zr.missing-sigma.replay-ppi-uniform.primary-delta\endcsname{0.004}
\expandafter\def\csname cnt@rev.pol.zr.missing-sigma.replay-ppi-uniform.primary-delta\endcsname{0}
\expandafter\def\csname pct@rev.pol.zr.missing-sigma.replay-ppi-uniform.primary-delta\endcsname{0.4}
\expandafter\def\csname ppt@rev.pol.zr.missing-sigma.replay-ppi-uniform.primary-delta\endcsname{$+$0.4}
\expandafter\def\csname val@rev.pol.zr.missing-sigma.uniform-fresh.n-changed\endcsname{82}
\expandafter\def\csname cnt@rev.pol.zr.missing-sigma.uniform-fresh.n-changed\endcsname{82}
\expandafter\def\csname pct@rev.pol.zr.missing-sigma.uniform-fresh.n-changed\endcsname{8200.0}
\expandafter\def\csname ppt@rev.pol.zr.missing-sigma.uniform-fresh.n-changed\endcsname{$+$8200.0}
\expandafter\def\csname val@rev.pol.zr.missing-sigma.uniform-fresh.n-changed-cat\endcsname{54}
\expandafter\def\csname cnt@rev.pol.zr.missing-sigma.uniform-fresh.n-changed-cat\endcsname{54}
\expandafter\def\csname pct@rev.pol.zr.missing-sigma.uniform-fresh.n-changed-cat\endcsname{5400.0}
\expandafter\def\csname ppt@rev.pol.zr.missing-sigma.uniform-fresh.n-changed-cat\endcsname{$+$5400.0}
\expandafter\def\csname val@rev.pol.zr.missing-sigma.uniform-fresh.primary-delta\endcsname{0.001}
\expandafter\def\csname cnt@rev.pol.zr.missing-sigma.uniform-fresh.primary-delta\endcsname{0}
\expandafter\def\csname pct@rev.pol.zr.missing-sigma.uniform-fresh.primary-delta\endcsname{0.1}
\expandafter\def\csname ppt@rev.pol.zr.missing-sigma.uniform-fresh.primary-delta\endcsname{$+$0.1}
\expandafter\def\csname val@rev.pol.zr.n-A\endcsname{4}
\expandafter\def\csname cnt@rev.pol.zr.n-A\endcsname{4}
\expandafter\def\csname pct@rev.pol.zr.n-A\endcsname{400.0}
\expandafter\def\csname ppt@rev.pol.zr.n-A\endcsname{$+$400.0}
\expandafter\def\csname val@rev.pol.zr.n-C\endcsname{12}
\expandafter\def\csname cnt@rev.pol.zr.n-C\endcsname{12}
\expandafter\def\csname pct@rev.pol.zr.n-C\endcsname{1200.0}
\expandafter\def\csname ppt@rev.pol.zr.n-C\endcsname{$+$1200.0}
\expandafter\def\csname val@rev.pol.zr.n-C-catalogue\endcsname{10}
\expandafter\def\csname cnt@rev.pol.zr.n-C-catalogue\endcsname{10}
\expandafter\def\csname pct@rev.pol.zr.n-C-catalogue\endcsname{1000.0}
\expandafter\def\csname ppt@rev.pol.zr.n-C-catalogue\endcsname{$+$1000.0}
\expandafter\def\csname val@rev.pol.zr.n-deferred\endcsname{16}
\expandafter\def\csname cnt@rev.pol.zr.n-deferred\endcsname{16}
\expandafter\def\csname pct@rev.pol.zr.n-deferred\endcsname{1600.0}
\expandafter\def\csname ppt@rev.pol.zr.n-deferred\endcsname{$+$1600.0}
\expandafter\def\csname val@rev.pol.zr.n-flagged\endcsname{30}
\expandafter\def\csname cnt@rev.pol.zr.n-flagged\endcsname{30}
\expandafter\def\csname pct@rev.pol.zr.n-flagged\endcsname{3000.0}
\expandafter\def\csname ppt@rev.pol.zr.n-flagged\endcsname{$+$3000.0}
\makeatother

\expandafter\def\csname val@rev.misc.zr.sigma-extra-changes\endcsname{1}
\expandafter\def\csname cnt@rev.misc.zr.sigma-extra-changes\endcsname{1}

\title{Visual Skill Recertification Across Model Changes}
\hypersetup{pdftitle={Visual Skill Recertification Across Model Changes},pdfauthor={Shengwei Zhang, Shenkang Qian, Haoda Dai, Yuheng Song, Boming Zhang}}
\begin{document}
\maketitle
\begin{abstract}
Visual skills, programs choosing where a vision-language model looks before clicking, need recertification after model changes. Treating skills as replayable evidence policies, we test replay (the new model re-reads stored views, one request per task) and budgeted re-execution for KEEP/SUSPEND decisions on two desktop grounding benchmarks, with every preregistered maintenance decision frozen before any sealed application was touched. Scope: three models of one provider, two tier changes toward a larger model, one harness, decoding configuration and image contract; cross-family migrations, downgrades and dated snapshots are untested. In this regime skill value changes sign across domains (\ppt{summary.sspro.tau-T.min} to \ppt{summary.sspro.tau-T.max} points on ScreenSpot-Pro, \ppt{summary.uiv.tau-T.min} to \ppt{summary.uiv.tau-T.max} on our UI-Vision subsample), yet no effect detectably flipped after either model change (post hoc): the comparisons cannot test the case recertification is for. Pooled over skills, replay is close to unbiased where its click agrees with the new model's direct click without drift (partly by that cell's construction; not per skill), and falls short mainly where the old model's zoomed views all missed the target (exploratory). Without flips, fresh-run correction of replay did not pay off: the stratified correction's preregistered equal-budget advantage is not shown beyond the realised pool (two-stage resampling: \cnt{summary.e6.n-ci-includes-0} of \cnt{summary.e6.n} intervals against competitors include zero; exploratory), is untested beyond our fixed-count allocation, mostly shrinks once history noise is removed (post hoc), and at exact equal cost does not consistently beat uniform re-evaluation (post hoc). Replay alone scored at least as high as every per-skill, fixed-fraction budgeted policy (post hoc): the old decision reached the sealed-test oracle in \cnt{summary.old-decision-at-oracle} of \cnt{summary.old-decision-at-oracle-n} settings while pool estimates transferred imperfectly; in the fourth, replay's pessimism suspended harmful skills.
\end{abstract}

\section{Introduction}
\label{sec:intro}
Vision-language agents that operate desktop software rarely look at a screenshot once. A grounding workflow decides where to look---a crop around a predicted centre, a $2{\times}2$ cell to expand, two candidate regions, an iterative zoom---renders the chosen views, and reads a click from them. Such workflows are now written down as reusable \emph{visual skills}---executable programs~\cite{xu2026viewcraft} or image--text and multimodal packages~\cite{jiang2026visualskill,zhang2026mmskills}---that are tuned on one model, stored, and loaded again on the next task. Their value is a property of the model that runs them. The benefit of test-time zooming depends on the model and the zoom configuration~\cite{jiang2025zoominclickout}, skill utility flips sign across models and harnesses~\cite{cheng2026agentskillsutility}, and causal attributions of skills have to be recomputed per model~\cite{wang2026notallskillshelp}. Models, however, change often: a move to another tier of a provider's family (the case we study), a dated snapshot, a migration to another provider. Each change raises the same maintenance question for every stored skill: is this skill still worth running on the new model, or should the agent fall back to direct grounding?

The honest answer is a paired re-evaluation on the new model~\cite{kevin2026aces}, and it is expensive: a visual skill costs two to four model requests per task, a catalogue has several skills, and the re-evaluation recurs at every change. The cheap alternative is \emph{replay}: the previous certification stored the evidence packets the old model produced---the crops and thumbnails it chose, with their geometry---so the new model can re-read each packet at one request per task. Replay fixes the evidence and changes only the reader, so the replay proxy $\rho$ differs from the gain $\tau$ the maintainer needs by a bias that depends on how differently the new model would have looked. Replay probes on fixed evidence are informative but not a correctness certificate~\cite{alhamidi2026carve}, and evidence and reader interact~\cite{xu2026viewcraft}. Prediction-powered inference~\cite{angelopoulos2023ppi,fisch2024stratppi} combines such a proxy with a few expensive measurements; what is open is which label-free features of a replay say where the proxy fails, and whether correcting it pays off for the decision.

We treat a visual skill as an evidence policy---an acquirer that asks the model for inspection regions, a CPU renderer that produces views under one image contract and records their geometry, and a reader that returns a click from the final views in a fresh session---so that a packet can be stored and replayed exactly (Sec.~\ref{sec:skills}), and define the gain $\tau$, proxy $\rho$ and bias $b$ on a maintenance pool (Sec.~\ref{sec:estimands}). \emph{Replay-corrected recertification} (RCR) replays every maintenance task, stratifies tasks by whether the replayed click agrees with the target model's direct click and whether it drifted from the source model's click, corrects the replay estimate with a budget of fresh executions allocated across strata, and decides KEEP or SUSPEND per skill (Sec.~\ref{sec:rcr}).

Decisions are frozen on a maintenance pool and scored on sealed, disjoint applications with a \emph{catalogue deployment score}---the accuracy an agent attains when it runs each kept skill and falls back to direct grounding otherwise---against keep-all, suspend-all, the old decision, replay-only, uniform fresh re-evaluation, replay-PPI, generic PPI, a best-single baseline and full recertification, at each policy's actual request cost (Sec.~\ref{sec:baselines}). The executors are three models of one provider; we study two tier changes toward a larger model. Every request is logged and audited, and protocols, decisions and analysis plan were frozen and hashed before any sealed-test request (Sec.~\ref{sec:setup}; Appendix~\ref{sup:amend}). Completed after the sealed results and marked as such: the Holm family and bootstrap $p$-values and seeds, the equal-cost comparison (interpolation preregistered, competitors chosen afterwards), the comparisons against replay-only and the old decision, and the re-run of a few failed sealed-test requests; a revision added post hoc analyses of the recorded outcomes with no new model request (Appendix~\ref{sup:revision}).

We find, first, that under our image contract the sign of a skill's value is set by the domain: five programs improve both target models on ScreenSpot-Pro by \ppt{summary.sspro.tau-T.min} to \ppt{summary.sspro.tau-T.max} points and hurt or do nothing on our UI-Vision subsample (\ppt{summary.uiv.tau-T.min} to \ppt{summary.uiv.tau-T.max}), where direct accuracy is high and the contract shrinks screenshots less, which may explain part of the contrast (Sec.~\ref{sec:exp_exec}). Within a domain no skill's effect changed sign with intervals excluding zero, our operational definition of a material flip (post hoc), although effects moved by up to \ppt{rev.desc.summary.d1.same-T.delta.max} points on the sealed test and \ppt{rev.desc.summary.d1.same-C.delta.max} on the maintenance pool; our data do not contain the case that motivates recertification. Second, pooled over skills, replay is close to unbiased on the \pct{summary.all.agree-stable-share.min}--\pct{summary.all.agree-stable-share.max}\% of tasks where the replayed click agrees with the target's direct click and has not drifted, but not per skill; an offline visual diagnostic (exploratory) locates its shortfall in the \pct{summary.expl.notcov-share.min}--\pct{summary.expl.notcov-share.max}\% of tasks where none of the source packet's zoomed views contained the target (Sec.~\ref{sec:exp_bias}). Third, without flips, correcting replay did not pay off. The stratified correction's preregistered equal-budget advantage (\cnt{holm.n-rejected} of \cnt{holm.family-size} Holm-corrected tests over 20 maintenance histories) is not shown beyond the realised pool, since resampling the pool gives intervals that include zero in \cnt{summary.e6.n-ci-includes-0} of \cnt{summary.e6.n} RCR-versus-competitor comparisons (exploratory); with \cnt{rev.pol.n-hist} histories \cnt{rev.pol.holm.n.n-rejected} reject (post hoc), and at equal request cost the advantage is not consistent. Fourth (post hoc), replay-only scored at least as high as every per-skill, fixed-fraction budgeted policy and an exploratory triggered policy: the old decision reached the sealed-test oracle in \cnt{summary.old-decision-at-oracle} of \cnt{summary.old-decision-at-oracle-n} settings while decisions estimated on $\mathcal{C}$ transferred imperfectly to new applications, and in the fourth, replay's pessimism suspended skills that hurt (Sec.~\ref{sec:exp_policies},~\ref{sec:exp_failures}).

Our contributions are (i) a formulation of visual skills as replayable evidence policies and of recertification after a model change as a budgeted decision problem with explicit estimands; (ii) a characterisation of replay bias by observable strata of the replayed click and by whether the source packet contained the target, which is what makes the problem visual: what replay misses is associated with where the old model looked (exploratory, Sec.~\ref{sec:discussion}); (iii) as a secondary result, a sealed-test comparison of per-skill, fixed-fraction recertification policies, with the negative result, largely post hoc, that none improved on replay when no skill materially flipped, a regime that cannot test the case recertification is for; and, as a supporting artefact, (iv) an audited set of direct, full-workflow and replay outcomes for five skills, three models and two domains with per-request provenance. The scope is two same-provider tier changes toward a larger model under one harness, decoding configuration and image contract; cross-family migrations, downgrades and dated snapshots are untested. The research artifact contains the code, requests, records and frozen plans (Appendix~\ref{sup:repro}).

\section{Related Work}
\label{sec:related}

\paragraph{Test-time visual search for GUI grounding.}
Guided visual search~\cite{wu2024vstar} made where-to-look a test-time decision for multimodal models; desktop grounding benchmarks~\cite{pmlr-v267-nayak25a,li2025screenspotpro} require localizing small elements in full-screen screenshots, and ScreenSpot-Pro~\cite{li2025screenspotpro} already reports cropping, iterative-narrowing and planner-guided baselines. Iterative narrowing~\cite{nguyen2024iterativenarrowing}, region-focused test-time scaling~\cite{luo2025regionfocus} and ZoomClick~\cite{jiang2025zoominclickout} refine the view in training-free rounds; ZoomClick improves every model it reports overall, but the benefit of depth and crop size is model-dependent and can turn negative for some models and samples. UI-Zoomer~\cite{tang2026uizoomer} gates zooming on token confidence combined with sampling consistency, and AutoFocus~\cite{yao2026autofocus} on a self-check followed by token perplexity. Our acquisition programs belong to this family; we propose no acquisition strategy and do not compare against these adaptive methods, which need token-level scores our harness does not expose or several sampled answers per task.

\paragraph{Executable visual skills and their transfer.}
ViewCraft~\cite{xu2026viewcraft} jointly evolves an executable evidence interface (crop, zoom, layout) and a reading program for a frozen executor, analyses evidence and reading transplants on fixed trajectories, and transfers saved skill programs across executors as-is. Its as-is transfer runs the program on the new executor, which corresponds to our natural workflow $F$, and its transplants exchange skill components within one frozen executor; replaying one executor's stored evidence packets on another executor, our $R$, is not part of it. Other work represents visual skills as image--text guides~\cite{jiang2026visualskill}, multimodal packages~\cite{zhang2026mmskills} or visual priors~\cite{xu2026visualskills}; VisualSkill reports in its limitations that on a weaker base model a skill's benefit can vanish~\cite{jiang2026visualskill}. These works construct, evaluate or transfer skills, but none decides how much target-side evaluation a transferred skill needs.

\paragraph{Skill evaluation after model changes.}
Skill utility is configuration-dependent: its sign flips across models and harnesses~\cite{cheng2026agentskillsutility}, randomized-mask causal attributions of skills are model-specific and recomputed per model~\cite{wang2026notallskillshelp}, and ACES evaluates skills by paired with- and without-skill trials with confidence intervals and recommends re-running them after model updates~\cite{kevin2026aces}. These studies motivate recertification and supply the paired protocol we adopt, but they either re-run the full evaluation or leave re-estimation after a model change open. Replay probes in CARVE~\cite{alhamidi2026carve} re-run fixed evidence through one frozen video pipeline and caution that replay sensitivity is not a correctness certificate, and reuse of unchanged visual inputs has been audited and certified within a single model~\cite{zeng2026visualorchestrationtax}; neither replays one model's evidence on another. CARVE also transfers a fitted decision rule across backbone sizes; we instead re-decide each stored skill from the new model's own measurements.

\paragraph{Model updates and regression testing.}
Regression-free model updates measure per-instance negative flips between an old and a new model and train the new model to reduce them~\cite{yan2021pct}; backward compatibility of updated models has been analysed for ML systems~\cite{srivastava2020backward} and human--AI teams~\cite{bansal2019updates}, and for language models as negative flips under base-model updates~\cite{echterhoff2024muscle} and as drift of a same-named service~\cite{chen2023chatgptdrift}. Selecting which tests to re-run after a change under a budget is the software-engineering form of the same problem~\cite{yoo2012regression}. We leave the model as it is and decide which stored skills to keep, measuring flips at the level of a skill's effect and, descriptively, of single tasks.

\paragraph{Proxy-assisted estimation.}
Prediction-powered inference~\cite{angelopoulos2023ppi} combines inexpensive predictions on many units with a small labelled sample to obtain valid confidence intervals, active statistical inference~\cite{zrnic2024active} directs the labelling budget toward uncertain units, and stratified PPI~\cite{fisch2024stratppi} combines the estimator with stratified sampling and allocation. Our corrected estimator is an instance of the latter; we adopt these estimators rather than contributing one. What we study is their use for budgeted recertification after a model change, with the target model's re-reading of the source model's evidence packets as the proxy.

\section{Problem and Method}
\label{sec:method}

\subsection{Visual skills as executable evidence policies}
\label{sec:skills}
A \emph{visual skill} $s=(\mathrm{acq}_s,\mathrm{render}_s,\mathrm{read})$ is a versioned program that turns a screenshot $I$ and an instruction $q$ into a click. The acquirer $\mathrm{acq}_s$ asks the model $M$ for an inspection region in the coordinates of the view it currently sees (a box, a $2{\times}2$ cell, two candidate boxes, or a point), the renderer crops and resizes on the CPU and records the affine map from each sent view back to $I$, and after at most three acquisition requests the reader $\mathrm{read}$ receives the resulting \emph{evidence packet} $P$---an ordered set of pixel views with their geometry and nothing else---in a fresh session and returns a view identifier and a local coordinate, which the CPU maps back to $I$. The source model's reasoning, intermediate clicks and correctness never enter the packet. One program from each of four acquisition families forms the catalogue $\mathcal{K}$ (Sec.~\ref{sec:setup}); a public zoom procedure is added as an external instance. All views obey one image contract (longest edge $\le 1024$\,px, $\le 2^{20}$ pixels, no upscaling, a $640$\,px global thumbnail where the family uses one), so differences between models are not confounded by provider-side resizing.

\subsection{Estimands}
\label{sec:estimands}
Fix a task distribution $\mathcal{Q}$ over (application, screenshot, instruction) and a scorer $y\in\{0,1\}$ (click inside the official box). For a model $m$ let $D_i=y(\mathrm{direct}(m,I_i,q_i))$ be the no-skill outcome and $F_{is}=y(\mathrm{read}(m,P^{m}_{is}))$ the outcome of the full workflow, where $P^{m}_{is}$ is the packet $m$ itself produces. The marginal value of $s$ on $m$ is $\tau_m(s)=\mathbb{E}[F_{is}-D_i]$. After a model change $M_0\!\to\!M_1$ the maintainer holds, for a maintenance pool $\mathcal{C}$, the source packets $P^{0}_{is}$ and source outcomes from the previous certification, and must decide $\mathrm{KEEP}$ or $\mathrm{SUSPEND}$ for each $s$ before deploying on unseen applications $\mathcal{T}$. The quantity that matters is $\tau_{1}(s)$ on $\mathcal{T}$; what is cheap to measure is the \emph{replay} outcome $R_{is}=y(\mathrm{read}(M_1,P^{0}_{is}))$, one reader request per task instead of a full acquisition. We write $\rho_s=\mathbb{E}[R_{is}-D_i]$ for the replay proxy of $\tau_1(s)$ and $b_s=\mathbb{E}[F_{is}-R_{is}]$ for its bias. Replay is a controlled intervention---$M_1$ reads evidence selected by $M_0$---and is informative only to the extent that $b_s$ is small or can be estimated; a positive $\rho_s$ alone certifies nothing~\cite{alhamidi2026carve}.

\subsection{Replay-corrected recertification}
\label{sec:rcr}
Given $\mathcal{C}$ with $N$ tasks, the method (RCR) spends one direct request and one replay request per task, then a budget $B$ of fresh full executions of $s$ on $M_1$ selected by stratified sampling, and estimates
\begin{equation}
\hat\tau_1(s)=\underbrace{\tfrac1N\textstyle\sum_{i}R_{is}}_{\text{replay}}+\underbrace{\textstyle\sum_{h}\tfrac{N_h}{N}\,\overline{(F-R)}_{h}}_{\text{stratified correction}}-\tfrac1N\textstyle\sum_i D_i,
\label{eq:rcr}
\end{equation}
where $h$ indexes strata of observable, label-free task features and $\overline{(F-R)}_h$ is the mean residual over the $n_h$ tasks sampled uniformly without replacement inside stratum $h$. With a single stratum Eq.~\eqref{eq:rcr} is the classical difference estimator, which is also the prediction-powered mean estimator~\cite{angelopoulos2023ppi}; the stratified form with Neyman-type allocation is standard survey sampling, implies no PPI-specific guarantee and is not a contribution. The decision is $\mathrm{KEEP}$ iff $\hat\tau_1(s)\ge 0$. Counts $n_h$ follow a fixed-count rule: reserve $\min(2,N_h)$ per stratum, then apportion the remainder proportionally to $N_h\sigma_h$ with deterministic largest remainders, where $\sigma_h$ is the residual standard deviation estimated on a disjoint development split (floor $0.10$); if $B$ cannot cover the minima the budget point is declared infeasible and the policy falls back to direct execution. The conditional finite-population variance of the correction, $\sum_h (N_h/N)^2(1-n_h/N_h)\,\mathrm{var}_h/n_h$, is reported as a diagnostic only; it excludes model randomness and application sampling.

\paragraph{Strata.}
The strata are fixed before the main study and frozen. Two features are available for every maintenance task at no extra cost: whether the replayed click agrees with the target's own direct click (within $2\%$ of the image diagonal), and whether the replayed click drifted from the source model's original click (more than $10\%$ of the diagonal). We use four cells---\emph{agree/stable}, \emph{agree/drift}, \emph{disagree}, and \emph{parse failure}. The rule and its cut-offs were selected on the UI-Vision haiku$\to$sonnet pilot, whose tasks overlap the main study's development split ($\cnt{overlap.pilot-uiv.A.n}$ of $\cnt{overlap.pilot-uiv.A.of}$ tasks) and maintenance pool ($\cnt{overlap.pilot-uiv.C.n}$ of $\cnt{overlap.pilot-uiv.C.of}$ tasks, $\cnt{overlap.pilot-uiv.C.apps}$ of $\cnt{overlap.pilot-uiv.C.of-apps}$ applications); there the replay residual was largest in the disagreement cell, where replay is right less often than the target's own workflow. Rows evaluated on that pool are therefore in-sample for the stratification, and the sealed-test rows are the out-of-sample check (Sec.~\ref{sec:exp_bias}).

\subsection{Maintenance baselines and deployment metric}
\label{sec:baselines}
All maintenance policies see the same pool $\mathcal{C}$, the same direct outcomes and the same source cache, and are charged every request they issue (direct, replay, acquisition, reader, retries). We compare: \emph{keep-all} and \emph{suspend-all}; \emph{old decision} (the source-model certification carried over); \emph{replay-only} ($\hat\tau=\hat\rho$); \emph{uniform fresh} (the single-skill form of current-model attribution~\cite{wang2026notallskillshelp,kevin2026aces}: $B$ uniformly sampled full executions, no replay); \emph{replay-PPI} (Eq.~\eqref{eq:rcr} with one stratum); \emph{generic PPI} (the same estimator with the source model's own outcome $S_{is}$ as the predictor, so that no target-side pixels are replayed); \emph{best-single} (choose one skill or direct for all tasks from the same fresh sample); and \emph{full recertification} (all $N$ fresh executions). Decisions are frozen on $\mathcal{C}$ and evaluated on the sealed applications $\mathcal{T}$ with the \emph{catalogue deployment score}: for each catalogue skill the per-application mean of $F_{is}$ if KEEP and $D_i$ if SUSPEND, averaged with equal weight over the catalogue skills and then over applications. This measures the maintenance decisions for workflows a user has already chosen to run; it is not an endorsement of running four skills at once. Because decisions are made on $\mathcal{C}$ and scored on other applications, the score measures estimation and the transfer of decisions across applications jointly; we therefore also report each policy's estimation error on $\mathcal{C}$ (Sec.~\ref{sec:exp_policies}).

\subsection{What replay can and cannot identify}
\label{sec:identification}
Replaying $P^{0}$ on $M_1$ fixes the evidence and varies the reader; the natural workflow varies both. The difference $F-R$ therefore mixes the value of $M_1$'s own acquisition with reader randomness, and crossing readers and packets describes these differences without identifying an internal perceptual mechanism. On a preregistered subset of $\mathcal{T}$ we add the reverse direction ($M_0$ reads $M_1$'s packet): against $M_0$'s direct click it describes what a packet adds for that reader, and against $M_0$'s own workflow (where $M_0$ was itself a target) it compares two packets under one reader; neither separates evidence from reader in general (descriptive, supplement). Random same-size crops and context-restored packets were dropped under the budget-reduction rule (Appendix~\ref{sup:amend}).

\section{Experiments}
\label{sec:experiments}

\subsection{Setup}
\label{sec:setup}
\paragraph{Tasks and splits.}
We use two desktop GUI grounding benchmarks with official point-in-box scoring: UI-Vision element grounding~\cite{pmlr-v267-nayak25a} (\cnt{census.uiv.applications} open-source applications; one functional instruction per element and at most two elements per screenshot, \cnt{census.uiv.roots-kept} element roots) and ScreenSpot-Pro~\cite{li2025screenspotpro} (\cnt{census.sspro.applications} application sets, i.e.\ professional applications plus operating-system sets, \cnt{census.sspro.tasks} items, from which a seeded pool of 16 items per application was preregistered; for ScreenSpot-Pro, ``application'' below means application set). The retained UI-Vision roots match the pool in element type and target size (Appendix~\ref{sup:revision}). A target's median shorter side is \cnt{rev.desc.d4.uiv.all.target-short-side-native.median}\,px natively on UI-Vision and \cnt{rev.desc.d4.sspro.all.target-short-side-native.median}\,px on ScreenSpot-Pro, but ScreenSpot-Pro's larger screenshots (median long edge \cnt{rev.desc.d4.sspro.all.long-edge.median} vs.\ \cnt{rev.desc.d4.uiv.all.long-edge.median}\,px) are scaled to \pct{rev.desc.d4.sspro.all.downscale-factor.median}\% vs.\ \pct{rev.desc.d4.uiv.all.downscale-factor.median}\% in the direct view, leaving \val{census.sspro.median-target-side-text} vs.\ \val{census.uiv.median-target-side-text}\,px. Applications are the independent unit. Before any model call, applications were assigned by a seeded, category-stratified draw to disjoint roles: $\mathcal{A}$ (development: $\sigma_h$), $\mathcal{C}$ (maintenance pool: all decisions), $\mathcal{T}$ (sealed deployment test), and a pilot-validation pool used only to size the study. After the budget reduction described below, UI-Vision uses \cnt{uiv.tasks.A.n-apps}/\cnt{uiv.tasks.C.n-apps}/\cnt{uiv.tasks.T.n-apps} applications and \cnt{uiv.tasks.A.n-tasks}/\cnt{uiv.tasks.C.n-tasks}/\cnt{uiv.tasks.T.n-tasks} tasks in $\mathcal{A}$/$\mathcal{C}$/$\mathcal{T}$, and ScreenSpot-Pro \cnt{sspro.tasks.A.n-apps}/\cnt{sspro.tasks.C.n-apps}/\cnt{sspro.tasks.T.n-apps} applications and \cnt{sspro.tasks.A.n-tasks}/\cnt{sspro.tasks.C.n-tasks}/\cnt{sspro.tasks.T.n-tasks} tasks. No screenshot or application appears in two roles, and no request touched a $\mathcal{T}$ application before the maintenance decisions were frozen and hashed. ScreenSpot-Pro, first planned as an external check, was promoted to a second main domain after UI-Vision development data showed every skill below direct grounding, to include a regime where skills help; the cross-domain contrast was thus anticipated from development data and confirmed, not discovered, on sealed applications.

\paragraph{Models and model changes.}
The executors are three current vision-language models of one provider, haiku, sonnet and opus (API-reported identifiers \texttt{claude-haiku-5-5}, \texttt{claude-sonnet-5-5}, \texttt{claude-opus-5-5}), under a fixed decoding configuration (reasoning effort \emph{low}; the API does not expose temperature). We study two model changes, haiku$\to$sonnet and sonnet$\to$opus. These are same-provider tier changes, not dated-snapshot upgrades of one model, and we therefore speak of \emph{transfer} rather than \emph{upgrade}. Every request is a fresh session, logged with the API-returned model identifier, request identifiers, token usage and image hashes (Sec.~\ref{sec:exp_audit}).

\paragraph{Skills.}
The catalogue contains one program per family: \emph{center} (one round; the window shrinks to 60\% around the predicted centre), \emph{grid} (up to three rounds of $2{\times}2$ cell selection with 15\% margin), \emph{context} (one centre round; the reader also receives a global thumbnail) and \emph{dual} (two candidate regions with one refinement round; the reader receives the thumbnail and both crops). A larger eight-program catalogue (two depths per family) and a 16-task-per-application maintenance pool were frozen initially. When the budget turned out far smaller than planned, a reduction rule was written after a usage-limit incident and before any sealed-test request; the applied reduction went beyond it (one program per family, 8 maintenance tasks per application, and no haiku workflow on $\mathcal{T}$), and the logged statement that no per-skill outcome was inspected first cannot be verified (Appendix~\ref{sup:amend}); the sealed test was cut to 6 (UI-Vision) and 8 (ScreenSpot-Pro) tasks per application under the rule. A fifth, external program, \emph{ZoomClick-noPZ}, is a ZoomClick-style narrowing, not ZoomClick itself: ratio 0.5 and 768\,px minimum crop as in~\cite{jiang2025zoominclickout}, but no five-call pre-zoom stage, two point-output rounds instead of three, a separate reader request and our contract; it is reported separately. Every program is capped at three acquisition requests plus one reader request; one format-correction retry is allowed and charged.

\paragraph{Protocol.}
For each task in $\mathcal{A}$ and $\mathcal{C}$ we run every skill on every model, direct grounding on every model, and replays of each source packet on the corresponding target; on $\mathcal{T}$ we run the target models' workflows, direct grounding (three independent runs per target model, which also yields a direct self-consistency baseline), and on a preregistered subset of $\mathcal{T}$ the reverse-direction readers of Sec.~\ref{sec:identification}. Maintenance policies are evaluated at fresh-execution budgets of 10, 25 (primary), 50 and 100\% of $|\mathcal{C}|$ per skill over 20 seeded maintenance histories (re-run offline with \cnt{rev.pol.n-hist} histories in a post hoc revision analysis); costs count every request, including replays and the direct requests shared by all policies. Uncertainty is reported with application-level paired bootstrap intervals over $\mathcal{T}$, conditional on the one maintenance pool; an exploratory leave-one-$\mathcal{C}$-application-out analysis probes that condition. The preregistered comparisons are RCR against replay-PPI and against generic PPI at equal fresh budget (one-sided), non-inferiority to full recertification with a one-point margin, and an equal-cost comparison by linear interpolation between budget points. Two of these were completed only after the sealed results had been computed: the Holm family (all \cnt{holm.family-size} one-sided superiority tests) and the bootstrap $p$-values were fixed after unadjusted lower bounds had been printed, and the equal-cost comparison was missing from our first analysis and added after an internal review; both follow the plan, which fixes the interpolation but not the competitors (we use every budgeted policy). Analyses beyond the plan are labelled exploratory or post hoc.

\subsection{Skill value changes sign across domains}
\label{sec:exp_exec}
The gain $\hat\tau=\overline{F-D}$ of every skill on the sealed applications is shown per skill, with intervals, in Figure~\ref{fig:skills}. On ScreenSpot-Pro, whose targets the contract shrinks more (Sec.~\ref{sec:setup}; possibly part of the reason, untested), every skill helps both target models: $\hat\tau$ ranges from \ppt{summary.sspro.p1.tau-T.min} to \ppt{summary.sspro.p1.tau-T.max} points for sonnet and from \ppt{summary.sspro.p2.tau-T.min} to \ppt{summary.sspro.p2.tau-T.max} for opus. On UI-Vision the same programs hurt sonnet (\ppt{summary.uiv.p1.tau-T.min} to \ppt{summary.uiv.p1.tau-T.max}) and are at best neutral for opus (\ppt{summary.uiv.p2.tau-T.min} to \ppt{summary.uiv.p2.tau-T.max}). Against the medoid of three direct runs the gains are \ppt{summary.sspro.gain-vs-medoid.min} to \ppt{summary.sspro.gain-vs-medoid.max} on ScreenSpot-Pro and \ppt{summary.uiv.gain-vs-medoid.min} to \ppt{summary.uiv.gain-vs-medoid.max} on UI-Vision, so the signs hold under our fixed image contract. Our direct accuracy is higher on UI-Vision (\pct{uiv.p1.catalog.suspend-all.score}--\pct{uiv.p2.catalog.suspend-all.score}\%) than on ScreenSpot-Pro (\pct{sspro.p1.catalog.suspend-all.score}--\pct{sspro.p2.catalog.suspend-all.score}\%), the reverse of the ordering published for earlier models~\cite{pmlr-v267-nayak25a,li2025screenspotpro}: our UI-Vision subsample leaves little headroom for zoom, and the sign contrast describes this regime, contract and model family. On UI-Vision the right decision is to suspend, on ScreenSpot-Pro to keep, and a maintenance policy must find out which from a small pool. We call a model change a \emph{material flip} for a skill if the skill's effect changes sign with application-level 95\% intervals excluding zero under both models (post hoc, as is everything from here to the end of this paragraph). No skill-setting meets this definition: none of \cnt{rev.desc.summary.d1.same-T.n} on $\mathcal{T}$ (sonnet$\to$opus) and none of \cnt{rev.desc.summary.d1.same-C.n} on $\mathcal{C}$ (both counts include the public skill), and none in the cross-pool haiku$\to$sonnet comparison (haiku's workflow on $\mathcal{T}$ was cancelled in the budget reduction; Appendix~\ref{sup:amend} and~\ref{sup:revision}). Effects did move: the cross-model change in $\tau$ reaches \ppt{rev.desc.summary.d1.same-T.delta.max} points on $\mathcal{T}$ (interval excluding zero for \cnt{rev.desc.summary.d1.same-T.delta-ci-excludes-0} skills) and \ppt{rev.desc.summary.d1.same-C.delta.max} on $\mathcal{C}$, equivalence within $\pm5$ points could not be established for any skill (TOST), and sealed effects below \pct{rev.pol.mdd-tau.uiv.min}--\pct{rev.pol.mdd-tau.uiv.max} points on UI-Vision are not detectable with 80\% power at one-sided $\alpha=0.05$, so ``no flip'' means no detectable sign change. Per task, sonnet$\to$opus workflow outcomes differ on up to \pct{rev.desc.summary.d2.uiv.p2.T.F.rate-any.max}\% of sealed tasks, against up to \pct{rev.desc.summary.d2.rep.T.rate-any.max}\% between repeated direct runs of one model (Table~\ref{tab:revflips}). With no sign change, the sealed-test-correct decision did not change, except at one exact-zero effect. On $\mathcal{C}$, old- and new-model estimates differ in sign for \cnt{summary.uiv.pool-sign-flips-with-public} of \cnt{summary.uiv.pool-sign-n-with-public} non-zero UI-Vision skill-settings (with the public skill), each at a one-task new-model estimate, none holding on $\mathcal{T}$ (Appendix~\ref{sup:estimators}).

\subsection{Where replay is biased, and what it misses}
\label{sec:exp_bias}
Pooling the catalogue skills, we split the replay residual $F-R$, the per-task contribution to the bias $b$, by the frozen strata (Table~\ref{tab:bias}). The agree/stable cell holds \pct{summary.all.agree-stable-share.min}--\pct{summary.all.agree-stable-share.max}\% of tasks, and its residual stays between \ppt{summary.all.agree-stable-resid.min} and \ppt{summary.all.agree-stable-resid.max} points: where the replayed click lands where the target model would click unaided, replay is, pooled over skills, close to unbiased on average, in a cell where both readers are mostly right. Per skill (Table~\ref{tab:revagree}) the residual spans \ppt{summary.all.agree-stable-resid-perskill.min} to \ppt{summary.all.agree-stable-resid-perskill.max} and none of \cnt{rev.desc.summary.d3.perskill-n} intervals excludes zero, but they reach \ppt{rev.desc.summary.d3.perskill-ci.lo-min} to \ppt{rev.desc.summary.d3.perskill-ci.hi-max} points, as wide as the effects that decide KEEP; and because the cell is defined by agreement with the target's direct click, part of this closeness follows from the definition. The disagreement cell holds \pct{summary.all.disagree-share.min}--\pct{summary.all.disagree-share.max}\% of tasks and its residual exceeds the agree/stable residual in all \cnt{summary.all.disagree-gap.n} settings. On the sealed applications, the out-of-sample check for the stratification, the gap is \ppt{expl.uiv.p2.e3.T.gap} [\ppt{expl.uiv.p2.e3.T.ci95.0}, \ppt{expl.uiv.p2.e3.T.ci95.1}] points on UI-Vision and \ppt{expl.sspro.p2.e3.T.gap} [\ppt{expl.sspro.p2.e3.T.ci95.0}, \ppt{expl.sspro.p2.e3.T.ci95.1}] on ScreenSpot-Pro (app-level cluster bootstrap; exploratory). On the maintenance pools the intervals are wide and \cnt{summary.e3.C.n-ci-includes-0} of \cnt{summary.e3.C.n} include zero (Appendix~\ref{sup:results}); the largest gap (UI-Vision haiku$\to$sonnet on $\mathcal{C}$) is in-sample for the strata selection (Sec.~\ref{sec:rcr}). The sealed rows exist for sonnet$\to$opus only, because the haiku workflow was not executed on $\mathcal{T}$.

\begin{figure}[!htbp]
\centering
\includegraphics[width=\linewidth]{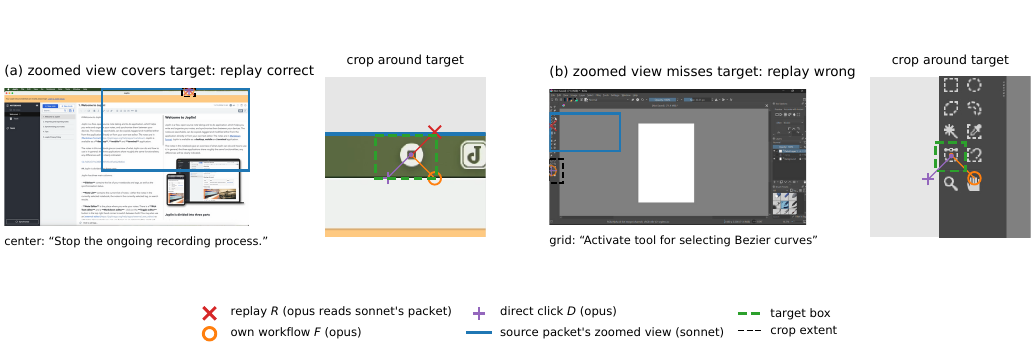}
\caption{Two sealed-test tasks (UI-Vision, sonnet$\to$opus), the first in identifier order over the single-view skills in which (a) the source packet's zoomed view covers the target and replay is correct, and (b) the view misses the target, replay is wrong and the target's own workflow is right; in (b) the target's direct click also falls outside the source view (the label-free signal of Table~\ref{tab:vision}). Markers in crops are offset; the line leads to the click. Screenshots from \href{https://huggingface.co/datasets/ServiceNow/ui-vision}{UI-Vision (MIT License)}, shown with crops and annotations.}
\label{fig:qual}
\end{figure}

\emph{What replay misses} (exploratory; Table~\ref{tab:vision}, Fig.~\ref{fig:qual}, Table~\ref{tab:visiondetail}). For \pct{summary.expl.notcov-share.min}--\pct{summary.expl.notcov-share.max}\% of tasks the target's centre lies outside every zoomed view of the source packet. Replay then reads single-view packets correctly only when the target box overlaps the view's edge (\pct{summary.expl.single-missed-accR.min}--\pct{summary.expl.single-missed-accR.max}\%) and thumbnail packets at thumbnail resolution (\pct{summary.expl.multi-missed-accR.min}--\pct{summary.expl.multi-missed-accR.max}\%); the target's own workflow, which can look elsewhere, reaches \pct{summary.expl.notcov-accF.min}--\pct{summary.expl.notcov-accF.max}\%. Where a zoomed view covers the target the residual is \ppt{summary.expl.covered-resid.min} to \ppt{summary.expl.covered-resid.max} points: replay still errs there (missed tasks hold \pct{summary.expl.err-share-missed.min}--\pct{summary.expl.err-share-missed.max}\% of its errors), but these errors are offset by tasks the target's workflow gets wrong, so replay's net shortfall comes mainly from the missed tasks. This diagnostic uses the answer keys; a label-free counterpart does not: whether the target model's own direct click falls outside the source packet's zoomed views. It lies almost entirely outside agree/stable (\pct{summary.expl.flag-outside-agree.min}--\pct{summary.expl.flag-outside-agree.max}\% of flagged tasks), and splitting the disagreement cell by it explains more residual variance than a random split only on the UI-Vision haiku$\to$sonnet pool, which overlaps the strata pilot ($p=\val{expl.uiv.p1.e10.C.perm-p}$, permutation within application, one of \cnt{summary.expl.perm-p.n} unadjusted tests); elsewhere, including both sealed rows, $p\ge\val{summary.expl.perm-p.other-min}$. The disagreement cell contains \pct{summary.expl.disagree-notcov.min}--\pct{summary.expl.disagree-notcov.max}\% missed targets against \pct{summary.expl.agree-notcov.min}--\pct{summary.expl.agree-notcov.max}\% in agree/stable. These are associations: crossing packets and readers does not identify a mechanism (Sec.~\ref{sec:identification}).

\begin{table}[!htbp]
\centering
\footnotesize
\setlength{\tabcolsep}{4pt}
\begin{adjustbox}{max width=\linewidth}\begin{tabular}{l r c r r r c c c c}
\toprule
 & \multicolumn{3}{c}{target missed by every zoomed view} & hit & errors & \multicolumn{2}{c}{miss \% in} & \multicolumn{2}{c}{label-free flag: direct click outside} \\
\cmidrule(lr){2-4}\cmidrule(lr){7-8}\cmidrule(lr){9-10}
Setting & \% & $R$/$F$ (\%) & $F{-}R$ & $F{-}R$ & in miss \% & agr./st. & disagr. & $N$ / $F{-}R$ & $+\eta^2$ / chance / $p$ \\
\midrule
UI-Vision, haiku$\to$sonnet, $\mathcal{C}$ & 18 & 7.5/44.8 & +37.3 & +0.3 & 55 & 8 & 49 & 45 / +44.4 & 7.0 / 1.8 / 0.030 \\
UI-Vision, sonnet$\to$opus, $\mathcal{C}$ & 9 & 8.6/31.4 & +22.9 & $-$0.9 & 39 & 5 & 37 & 20 / +15.0 & 2.0 / 2.9 / 0.827 \\
UI-Vision, sonnet$\to$opus, $\mathcal{T}$ & 8 & 8.9/22.2 & +13.3 & 0.0 & 39 & 5 & 36 & 19 / +31.6 & 2.0 / 0.7 / 0.168 \\
ScreenSpot-Pro, haiku$\to$sonnet, $\mathcal{C}$ & 14 & 8.3/20.8 & +12.5 & $-$2.0 & 73 & 10 & 30 & 10 / 0.0 & 0.2 / 1.5 / 1.000 \\
ScreenSpot-Pro, sonnet$\to$opus, $\mathcal{C}$ & 14 & 3.7/29.6 & +25.9 & $-$3.1 & 72 & 5 & 61 & 21 / +19.0 & 1.0 / 1.5 / 0.716 \\
ScreenSpot-Pro, sonnet$\to$opus, $\mathcal{T}$ & 4 & 0.0/28.6 & +28.6 & +0.6 & 36 & 1 & 27 & 17 / +5.9 & 0.1 / 0.6 / 1.000 \\
\bottomrule
\end{tabular}\end{adjustbox}

\caption{Exploratory visual diagnostic (catalogue skills pooled; rows without a completed source packet excluded): $R$/$F$ accuracy and residual $F-R$ (points) when the target centre is outside every zoomed source view (miss) or inside one (hit); share of replay errors on misses; share of misses per stratum; label-free flag (target direct click outside the zoomed views: $N$, mean $F-R$); last column: $\eta^2$ gained (points, ${\ge}0$ by construction) by splitting the disagreement cell by the flag, its mean under random within-application splits (chance) and the permutation $p$.}

\label{tab:vision}
\end{table}

\subsection{Maintenance policies on sealed applications}
\label{sec:exp_policies}
Without material flips these policy comparisons are secondary: they cannot test the case recertification is for (Sec.~\ref{sec:exp_exec}). Table~\ref{tab:policies} reports the catalogue deployment score on $\mathcal{T}$ for every policy, with its KEEP rate and request cost on $\mathcal{C}$; Table~\ref{tab:equalcost} gives the equal-cost comparison, and Figure~\ref{fig:cost} traces score against cost over all budgets.

\emph{Estimation on $\mathcal{C}$} (exploratory). At equal fresh budget RCR has the lowest error of $\hat\tau$ against the full-pool estimate in \cnt{summary.expl.rcr-lowest-rmse} of \cnt{summary.expl.rcr-rmse-settings} settings, but it also spends one replay per task; at RCR's request cost, uniform fresh or generic PPI is more accurate in all \cnt{summary.expl.eqcost-simpler-more-accurate} settings, and the uncorrected replay estimate, which needs no fresh execution, is more accurate than RCR in \cnt{summary.expl.replay-only-lower-rmse-than-rcr} of \cnt{summary.expl.replay-only-rmse-n} (Table~\ref{tab:estimator}).

\emph{Equal fresh budget} (preregistered). Over the 20 frozen histories RCR scores higher than replay-PPI and generic PPI in all \cnt{holm.family-size} comparisons, by \ppt{summary.all.primary-gain.min} to \ppt{summary.all.primary-gain.max} points, and after Holm correction \cnt{holm.n-rejected} reject at $\alpha=0.05$. These means carry Monte Carlo error from the 20 histories (standard error \pct{rev.pol.holm.mc.min-se20}--\pct{rev.pol.holm.mc.max-se20} points), as large as the differences. Post hoc, with \cnt{rev.pol.n-hist} histories \cnt{rev.pol.holm.n.n-rejected} of \cnt{holm.family-size} reject (\cnt{rev.pol.holm.n-rejected-both} of the frozen \cnt{holm.n-rejected} survive; both ScreenSpot-Pro sonnet$\to$opus differences fall to about zero), no rejection depends on bootstrap size or seed, and sign-flip tests over applications reject \cnt{rev.pol.perm.n.signflip-gt.n-rejected} (Tables~\ref{tab:revholm}--\ref{tab:revtests}). At 80\% power these pools detect differences of \pct{rev.pol.mdd.f20.sup.min}--\pct{rev.pol.mdd.f20.sup.max} points unadjusted and \pct{rev.pol.mdd.f20.sup.holm-min}--\pct{rev.pol.mdd.f20.sup.holm-max} at the Holm level, ignoring history noise (post hoc). Where the sealed oracle is the same for every skill, score differences arise only through which skills are kept, all in one direction (correlation over the \cnt{rev.pol.keep.uiv.p1.corr.budgeted-points} budget points of Table~\ref{tab:revkeep}: $\val{rev.pol.keep.uiv.p1.corr.pearson}$, $\val{rev.pol.keep.sspro.p1.corr.pearson}$, $\val{rev.pol.keep.sspro.p2.corr.pearson}$), so the score alone cannot separate better estimation from a directional bias. At the primary budget RCR's KEEP rate is the closest to the oracle in both haiku$\to$sonnet settings but not on ScreenSpot-Pro sonnet$\to$opus (Appendix~\ref{sup:revision}), so better decisions at equal fresh budget are possible but not established (post hoc). All of this is conditional on the realised maintenance pool: resampling maintenance and sealed applications together (exploratory) gives intervals that include zero in \cnt{summary.e6.n-ci-includes-0} of \cnt{summary.e6.n} RCR-versus-competitor comparisons. RCR is non-inferior to full recertification with a one-point margin in \cnt{summary.all.ni-pass} of \cnt{summary.all.ni-settings} settings at \pct{summary.all.rcr-cost-fraction.min}--\pct{summary.all.rcr-cost-fraction.max}\% of its requests.

\emph{Equal request cost} (interpolation preregistered, competitors chosen after the sealed results). RCR spends its fresh budget plus one replay per task, so at the same fresh budget it costs more than uniform fresh or generic PPI. Interpolated from the 20 frozen histories to RCR's cost (Table~\ref{tab:equalcost}), uniform fresh scores within \ppt{summary.eqcost.uniform-fresh.min} to \ppt{summary.eqcost.uniform-fresh.max} points of RCR (RCR minus competitor) and generic PPI within \ppt{summary.eqcost.generic-ppi-source.min} to \ppt{summary.eqcost.generic-ppi-source.max}, and RCR is above replay-PPI in every setting. Matching the cost exactly with \cnt{rev.pol.n-hist} histories (post hoc; Table~\ref{tab:reveqcost}) changes this: RCR is ahead of uniform fresh on UI-Vision haiku$\to$sonnet (\ppt{rev.pol.eqc.uiv.p1.uniform-fresh.minus.mean} [\ppt{rev.pol.eqc.uiv.p1.uniform-fresh.minus.ci95.0}, \ppt{rev.pol.eqc.uiv.p1.uniform-fresh.minus.ci95.1}]) but behind it in \cnt{rev.pol.eqc.uniform-fresh.n-ci-below-0} settings, behind generic PPI in \cnt{rev.pol.eqc.generic-ppi-source.n-ci-below-0}, and ahead of replay-PPI in \cnt{rev.pol.eqc.replay-ppi-uniform.n-ci-above-0} of \cnt{summary.all.ni-settings}.

\emph{No fresh execution at all} (post hoc comparisons). Replay-only scored at least as high as every per-skill budgeted policy, RCR included, in \cnt{summary.replay-only-ge-perskill-all-budgets} of \cnt{summary.replay-only-ge-perskill-budgeted-n} settings at every budget point, with or without the infeasible ones (Figure~\ref{fig:cost}; on ScreenSpot-Pro as a tie); suspend-all exceeds it in \cnt{summary.suspend-all-above-replay} (UI-Vision sonnet$\to$opus) and best-single, a different deployment, exceeds it on ScreenSpot-Pro sonnet$\to$opus; RCR minus replay-only is \ppt{summary.rcr-vs-replay-only.min} to \ppt{summary.rcr-vs-replay-only.max} points (interval below zero in \cnt{summary.rcr-vs-replay-only.n-ci-below-0} settings conditional on the pool), but the two-stage bootstrap puts this difference between \ppt{summary.e6.replay-only.min} and \ppt{summary.e6.replay-only.max} points with every interval including zero. The main reason is that the sealed-test-correct decision did not change (Sec.~\ref{sec:exp_exec}): the old decision attains the oracle in \cnt{summary.old-decision-at-oracle} of \cnt{summary.old-decision-at-oracle-n} settings, where replay-only coincides with it, while decisions estimated on $\mathcal{C}$ transfer imperfectly (Sec.~\ref{sec:exp_failures}); in the fourth (UI-Vision sonnet$\to$opus) the old decision kept skills that hurt on $\mathcal{T}$, and replay's pessimism (its estimate is below the full-pool one for \cnt{summary.expl.replay-pessimistic} of \cnt{summary.expl.replay-pessimistic-n} catalogue skill-settings) explains replay-only's score; the deployment score cannot reward estimation accuracy on $\mathcal{C}$ in these settings. Only per-skill, fixed-fraction policies were preregistered; an exploratory triggered policy that spends the same fresh budget only on skills whose replay estimate lies within two standard errors of zero scores \ppt{rev.pol.trig.se.minus-replay-only.min} to \ppt{rev.pol.trig.se.minus-replay-only.max} points relative to replay-only.

\begin{table}[!htbp]
\centering
\scriptsize
\setlength{\tabcolsep}{2.6pt}
\begin{adjustbox}{max width=\linewidth}\begin{tabular}{l rrrr rrrr rrrr rrrr}
\toprule
Policy (budget 25\% where applicable) & \multicolumn{4}{c}{\shortstack{UI-Vision\\haiku$\to$sonnet}} & \multicolumn{4}{c}{\shortstack{UI-Vision\\sonnet$\to$opus}} & \multicolumn{4}{c}{\shortstack{ScreenSpot-Pro\\haiku$\to$sonnet}} & \multicolumn{4}{c}{\shortstack{ScreenSpot-Pro\\sonnet$\to$opus}} \\
 & score & $N$ & K & req. & score & $N$ & K & req. & score & $N$ & K & req. & score & $N$ & K & req. \\
\midrule
suspend-all (direct) & 83.3 & -- & 0 & -- & 84.0 & -- & 0 & -- & 74.0 & -- & 0 & -- & 76.0 & -- & 0 & -- \\
keep-all & 76.0 & -- & 100 & -- & 81.6 & -- & 100 & -- & 85.9 & -- & 100 & -- & 90.6 & -- & 100 & -- \\
old decision & 83.3 & -- & 0 & -- & 81.8 & -- & 50 & -- & 85.9 & -- & 100 & -- & 90.6 & -- & 100 & -- \\
replay-only & 83.3 & -- & 0 & 192 & 83.3 & -- & 50 & 192 & 85.9 & -- & 100 & 96 & 90.6 & -- & 100 & 96 \\
\midrule
uniform fresh & 79.7 (2.5) & 79.2 & 57 & 161 & 82.1 (0.9) & 82.4 & 72 & 160 & 85.0 (2.2) & 85.3 & 95 & 81 & 89.1 (2.5) & 89.5 & 93 & 81 \\
generic PPI (source) & 79.0 (2.2) & 79.7 & 50 & 161 & 82.6 (0.9) & 82.7 & 54 & 160 & 85.2 (1.4) & 84.7 & 90 & 81 & 88.5 (1.9) & 89.1 & 90 & 81 \\
replay-PPI & 80.2 (2.3) & 80.0 & 44 & 257 & 82.8 (0.8) & 82.7 & 58 & 256 & 84.9 (1.6) & 85.2 & 95 & 129 & 88.5 (2.3) & 89.1 & 90 & 129 \\
RCR (stratified replay corr.) & 80.4 (1.9) & 80.5 & 38 & 257 & 82.9 (0.9) & 82.8 & 59 & 256 & 85.8 (0.8) & 85.6 & 97 & 129 & 89.5 (1.8) & 89.1 & 90 & 129 \\
\midrule
full recertification & 79.3 & -- & 50 & 356 & 83.2 & -- & 75 & 350 & 85.9 & -- & 100 & 180 & 90.6 & -- & 100 & 179 \\
catalogue oracle on $\mathcal{T}$ & 83.3 & -- & 0 & -- & 84.0 & -- & 25 & -- & 85.9 & -- & 100 & -- & 90.6 & -- & 100 & -- \\
\midrule
best-single (one skill for all tasks) & 80.2 (3.3) & 79.0 & -- & 161 & 83.1 (1.2) & 82.6 & -- & 160 & 85.3 (1.6) & 83.9 & -- & 81 & 91.3 (0.8) & 90.6 & -- & 81 \\
\bottomrule
\end{tabular}\end{adjustbox}

\caption{Catalogue deployment score (\%) on $\mathcal{T}$ and mean maintenance requests per skill on $\mathcal{C}$. Budgeted policies: 25\% of $|\mathcal{C}|$ fresh executions per skill, mean (SD) over the 20 frozen histories and ($N$, post hoc) over \cnt{rev.pol.n-hist}; K: share of catalogue skills kept (\%; budgeted policies over the $N$ histories). Best-single runs one skill or direct for every task, a different estimand that can exceed the catalogue oracle.}
\label{tab:policies}
\end{table}

\subsection{Failure cases}
\label{sec:exp_failures}
\emph{Decisions on $\mathcal{C}$ do not always transfer to new applications.} On UI-Vision haiku$\to$sonnet, full recertification itself falls short of suspend-all (difference \ppt{summary.uiv.p1.full-recert.vs-suspend-all} points). On $\mathcal{C}$ the center skill has $\hat\tau=$ \ppt{uiv.p1.decide.center-d1.no-budget-tau.full-recert} and the dual skill \ppt{uiv.p1.decide.dual-d2.no-budget-tau.full-recert} points, both kept under the frozen rule KEEP iff $\hat\tau\ge0$, while on $\mathcal{T}$ they cost sonnet \ppt{uiv.p1.execution-layer.center-d1.tau-vs-direct.mean} and \ppt{uiv.p1.execution-layer.dual-d2.tau-vs-direct.mean} points. When true effects are within a few points of zero, the pool's applications rather than the estimator limit the decision. Post hoc, a margin of up to two points, or one tied to a skill's extra requests per task, changes UI-Vision scores by up to \ppt{rev.pol.rules.gain.uiv.m2.max} points and ScreenSpot-Pro scores by down to \ppt{rev.pol.rules.gain.sspro.m2.min}, so no fixed margin helps everywhere (Appendix~\ref{sup:revision}, with the deployment requests each decision commits to); a larger, more diverse pool is untested.

\emph{Small pools make stratified allocation infeasible.} At the 10\% budget the ScreenSpot-Pro pool gives each skill too few fresh executions for two per stratum. On haiku$\to$sonnet this holds for every skill, RCR falls back to direct execution as preregistered and the catalogue score drops to \pct{sspro.p1.catalog.rcr-stratified@0.1.mean-over-histories}\%, the suspend-all score; on sonnet$\to$opus it holds for the grid skill only and the score drops to \pct{sspro.p2.catalog.rcr-stratified@0.1.mean-over-histories}\% (Figure~\ref{fig:cost}). A fallback to replay-PPI would avoid this; we report the frozen behaviour.

\subsection{Execution audit}
\label{sec:exp_audit}
The main study completed \cnt{uiv.records.ok-requests} (UI-Vision) and \cnt{sspro.records.ok-requests} (ScreenSpot-Pro) requests; the API-returned model identifier matched the requested model in all of them, and no actor read an existing file outside its request's allowed views. The \cnt{uiv.records.audit.requests-with-violations} and \cnt{sspro.records.audit.requests-with-violations} requests with an unexpected tool call gave the model no information beyond its request (Appendix~\ref{sup:results}). In one short window the smallest model returned \cnt{d020.n-total} answers without reading any file; re-running the \cnt{d020.n-T} sealed-test ones after the sealed results had been computed is a protocol deviation, and the \cnt{d020.n-C} that feed the frozen maintenance decisions remain in place; treating them as missing or wrong moves no catalogue score by more than \pct{rev.pol.zr.missing.max-abs-score-delta} points (Appendix~\ref{sup:results} and~\ref{sup:revision}).

\section{Discussion and Conclusion}
\label{sec:discussion}
\paragraph{What the evidence supports.} Under one image contract a skill's value can flip sign across benchmarks but did not detectably do so across our model changes. Pooled over skills, replay is close to unbiased in the agree/stable cell, partly by construction, and not reliably per skill.

\paragraph{Why recertification is a vision problem.} A generic proxy such as the old model's score says nothing about where the new model would look; replay re-reads the views the old model chose, and is right mainly when they contain the target (Sec.~\ref{sec:exp_bias}). Its bias largely reflects visual acquisition (plus reader randomness, Sec.~\ref{sec:identification}), and its label-free strata are geometric (an exploratory association, not a mechanism).

\paragraph{What it does not support.} Per-skill decisions from the agree/stable cell alone, or a ranking of recertification policies. Without material flips the old decision reached the oracle in \cnt{summary.old-decision-at-oracle} of \cnt{summary.old-decision-at-oracle-n} settings and decisions estimated on $\mathcal{C}$ transferred imperfectly to $\mathcal{T}$; in the fourth, replay's pessimism happened to suspend skills that hurt. The equal-budget advantage of stratified correction mostly shrinks once history noise is removed (post hoc), and at equal cost it is not a better estimator than uniform re-evaluation, conditional on the pool and our fixed-count allocation. Effects below the minimum detectable differences cannot be excluded, so it is also a statement about power. We do not recommend replay as a general decision rule.

\paragraph{Scope and limits.} After a downgrade the packets come from a stronger acquirer, so replay may over-estimate a skill (reverse readers, Table~\ref{tab:reverse}, descriptive). Decisions rest on \cnt{uiv.tasks.C.n-apps} and \cnt{sspro.tasks.C.n-apps} maintenance and \cnt{uiv.p1.n-test-apps} and \cnt{sspro.p1.n-test-apps} sealed applications, and intervals over so few clusters may under-cover. Hierarchical or sequential decisions, active allocation~\cite{zrnic2024active} and a pessimism adjustment of replay are untested.

\paragraph{Conclusion.} Replay errs mainly where the old model's views missed the target (exploratory). Without material flips no per-skill, fixed-fraction policy beat replay alone (post hoc), correction did not pay off at equal cost under our allocation (post hoc), and such comparisons cannot test the case recertification is for.

\section*{LLM Usage Statement}
The authors report that they wrote the research code and the manuscript draft and performed the analyses used in this version. For these tasks, AI assistance was limited to running existing scripts. The Claude vision-language models described in the experimental setup were systems under evaluation. Codex (OpenAI) assisted with preprint formatting and editing this disclosure. The authors take responsibility for the final work.

\bibliographystyle{unsrtnat}
\bibliography{references}

\clearpage
\appendix
\setcounter{figure}{0}
\renewcommand{\thefigure}{S\arabic{figure}}
\renewcommand{\theHfigure}{supp.\arabic{figure}}
\setcounter{table}{0}
\renewcommand{\thetable}{S\arabic{table}}
\renewcommand{\theHtable}{supp.\arabic{table}}
\section*{Appendices}
\section{Execution environment and model access}
\label{sup:env}
All model requests were issued through a coding-agent harness rather than a metered API: each request is a fresh, read-only agent session on one of three vision-language models of a single provider (haiku, sonnet, opus; API-reported identifiers \texttt{claude-haiku-5-5}, \texttt{claude-sonnet-5-5}, \texttt{claude-opus-5-5}), with reasoning effort fixed to \emph{low}. The agent can only read the request file and the image files it lists; an automated audit of every transcript checks the tool calls (reads outside the allowed files: none in the study; a handful of harmless harness-interface calls and answer echoes are logged), the API-returned model identifier (always equal to the requested alias), the provider request identifiers, token usage and the byte identity of every image the model received. The harness re-encodes large PNG files to JPEG before the model sees them; after this was detected in the pilot, every view whose PNG exceeds 480\,KB is encoded by us as a deterministic JPEG (quality ladder from 90 down, 4:4:4 subsampling), so that the file hash recorded is exactly what the model received. Because no dollar price is attached to the harness, decisions and cost curves use request counts; harness API calls, tokens, images and pixels are logged per request, and pixels sent per task are summarised in Table~\ref{tab:revpixels}. We make no claim about dollar savings.

\section{Image contract, skills and prompts}
\label{sup:skills}
Longest edge $\le 1024$\,px and $\le 2^{20}$ pixels per view, LANCZOS downscaling, no upscaling, PNG (or deterministic JPEG, Sec.~\ref{sup:env}); global thumbnails use a 640\,px longest edge; local windows never shrink below 128 original pixels on a side. \emph{center}: the model returns a box in the current view; the next window is $0.6\times$ the previous window's width and height centred on the box centre, clamped to the image. \emph{grid}: the current window is split into a $2{\times}2$ grid, the model names a cell, the cell is expanded by 15\% of its side on each side. \emph{context}: centre acquisition; the reader receives the thumbnail and the final local view. \emph{dual}: the model returns two boxes with IoU $<0.5$, each expanded by 15\%; one refinement round returns one tightened box per candidate; the reader receives the thumbnail and both final crops; if no valid pair is produced the packet is the thumbnail plus the full view. \emph{ZoomClick-noPZ}: the public ZoomClick narrowing (ratio 0.5 of the current window, minimum crop 768\,px, clip boundary mode, no history between rounds), implemented under our contract without the five-call pre-zoom stage, with two point-output acquisition rounds and a reader on the final crop. Readers return a view identifier and a local coordinate; coordinates outside the chosen view are scored as format failures, never clamped. One format-correction retry per job is allowed and charged. Full prompt texts and JSON schemas are in the research artifact.

\section{Protocol freeze and budget reduction}
\label{sup:amend}
The main-study protocols were frozen (SHA-256 recorded in the decision log) before any request on the sealed test applications. During the first main-study waves the execution budget of the host turned out to be far smaller than the frozen scale required: most requests of the first concurrent waves failed with a usage-limit error (logged as infrastructure failures, re-queued unchanged and excluded from all counts; Sec.~\ref{sup:infra}). After this incident, and before any sealed-test request, a reduction rule was written into the decision log (D-012: drop fixed-packet controls, then reduce sealed-test tasks per application, then direct repeats) and applied as an amended protocol (D-013). The applied reduction went beyond that written list: the catalogue was restricted to one program per family, the set already used in the pilot, plus the public skill; the maintenance pool to 8 tasks per application; and the planned haiku workflow on the reverse-reader subset of $\mathcal{T}$ was cancelled with the other E4 controls (D-013), which makes every haiku$\to$sonnet flip comparison cross-pool. Retained tasks are prefixes of the seeded task order. The partially collected data for the dropped programs are kept in the research artifact as descriptive material and enter no comparison. The amendment was made after part of the development and maintenance requests had been harvested; the decision log asserts that no per-skill main-study outcomes were inspected, which cannot be verified from the request logs, so we do not rely on it. The plan as it stood at the ScreenSpot-Pro freeze is stored as a separate file and re-derives that protocol's decision files byte for byte.

\section{Estimators and decision rules}
\label{sup:estimators}
\paragraph{Frozen analysis plan.}
All rules below are fixed in a machine-readable plan whose SHA-256 is recorded in the decision log before the first sealed-test request of each domain. Budget fractions are $\{10,25,50,100\}\%$ of $|\mathcal{C}|$ per skill with $25\%$ preregistered as primary; every budgeted policy is run over \val{plan.n-histories} seeded maintenance histories (the seed fixes which tasks are sampled); application-level paired bootstrap uses \cnt{plan.bootstrap-B} resamples. The decision rule is $\mathrm{KEEP}$ iff $\hat\tau\ge0$; an infeasible stratified allocation (budget below the per-stratum minima) is recorded and falls back to $\mathrm{SUSPEND}$, i.e.\ direct execution.

\paragraph{Strata and $\sigma_h$.}
Strata use two label-free features of the replayed click: agreement with the target model's own direct click (within \pct{plan.agree-cut}\% of the image diagonal) and drift from the source model's original click (more than \pct{plan.drift-cut}\% of the diagonal); parse failures form their own cell. The residual standard deviations $\sigma_h$ that drive the allocation were estimated per domain, model pair, skill and stratum on the development applications $\mathcal{A}$ only, with the default \val{plan.sigma-default} for cells with fewer than two development tasks and a floor of \val{plan.sigma-floor}; Table~\ref{tab:strata} lists the counts and the values used. Allocation reserves $\min(2,N_h)$ per stratum and apportions the rest proportionally to $N_h\sigma_h$ by largest remainders; sampling inside a stratum is uniform without replacement and the correction is the stratum-weighted mean residual (Eq.~\eqref{eq:rcr}). The strata rule and its cut-offs were selected on the UI-Vision haiku$\to$sonnet pilot, whose tasks overlap the main study's $\mathcal{A}$ (\cnt{overlap.pilot-uiv.A.n} of \cnt{overlap.pilot-uiv.A.of} tasks) and $\mathcal{C}$ (\cnt{overlap.pilot-uiv.C.n} of \cnt{overlap.pilot-uiv.C.of} tasks, \cnt{overlap.pilot-uiv.C.apps} of \cnt{overlap.pilot-uiv.C.of-apps} applications) for that setting (decision D-009 of the project decision log).

\begin{table}[!htbp]\centering\footnotesize
\caption{Strata on the maintenance pool $\mathcal{C}$ ($N_h$), development counts on $\mathcal{A}$ ($n_h$) and the $\sigma_h$ values used by the allocation, per domain, pair and skill.}
\label{tab:strata}
\begin{adjustbox}{max width=\linewidth}\begin{tabular}{l r cccc}
\toprule
Skill & $N$ & agree/stable & agree/drift & disagree & parse fail \\
 & & \multicolumn{4}{c}{$N_h$ on $\mathcal{C}$ / $n^{\mathcal{A}}_h$ on $\mathcal{A}$ / $\sigma_h$ used} \\
\midrule
\multicolumn{6}{l}{\emph{UI-Vision, haiku$\to$sonnet}} \\
\quad center & 96 & 70 / 52 / 0.14 & 7 / 2 / 0.50 & 17 / 9 / 0.50 & 2 / 1 / 0.50 \\
\quad grid & 96 & 62 / 45 / 0.21 & 8 / 5 / 0.10 & 22 / 11 / 0.50 & 4 / 3 / 0.47 \\
\quad context & 96 & 73 / 54 / 0.10 & 11 / 0 / 0.50 & 11 / 10 / 0.46 & 1 / 0 / 0.50 \\
\quad dual & 96 & 66 / 49 / 0.20 & 10 / 3 / 0.10 & 20 / 12 / 0.64 & 0 / 0 / -- \\
\quad ZoomClick-noPZ (public) & 96 & 76 / 49 / 0.10 & 6 / 3 / 0.10 & 13 / 12 / 0.72 & 1 / 0 / 0.50 \\
\addlinespace
\multicolumn{6}{l}{\emph{UI-Vision, sonnet$\to$opus}} \\
\quad center & 96 & 87 / 55 / 0.19 & 1 / 2 / 0.10 & 8 / 7 / 0.35 & 0 / 0 / -- \\
\quad grid & 96 & 77 / 52 / 0.10 & 2 / 0 / 0.50 & 15 / 11 / 0.51 & 2 / 1 / 0.50 \\
\quad context & 96 & 84 / 56 / 0.13 & 2 / 3 / 0.10 & 10 / 5 / 0.49 & 0 / 0 / -- \\
\quad dual & 96 & 86 / 54 / 0.13 & 4 / 1 / 0.50 & 6 / 9 / 0.47 & 0 / 0 / -- \\
\quad ZoomClick-noPZ (public) & 96 & 85 / 55 / 0.19 & 0 / 1 / -- & 11 / 8 / 0.60 & 0 / 0 / -- \\
\addlinespace
\multicolumn{6}{l}{\emph{ScreenSpot-Pro, haiku$\to$sonnet}} \\
\quad center & 48 & 38 / 26 / 0.10 & 1 / 0 / 0.50 & 8 / 3 / 0.82 & 1 / 3 / 0.47 \\
\quad grid & 48 & 35 / 25 / 0.20 & 0 / 0 / -- & 11 / 4 / 0.43 & 2 / 3 / 0.47 \\
\quad context & 48 & 36 / 26 / 0.19 & 5 / 2 / 0.50 & 6 / 4 / 0.43 & 1 / 0 / 0.50 \\
\quad dual & 48 & 37 / 27 / 0.42 & 1 / 1 / 0.50 & 8 / 4 / 0.43 & 2 / 0 / 0.50 \\
\quad ZoomClick-noPZ (public) & 48 & 38 / 25 / 0.10 & 2 / 1 / 0.50 & 7 / 6 / 0.75 & 1 / 0 / 0.50 \\
\addlinespace
\multicolumn{6}{l}{\emph{ScreenSpot-Pro, sonnet$\to$opus}} \\
\quad center & 48 & 42 / 29 / 0.10 & 0 / 0 / -- & 6 / 2 / 0.10 & 0 / 1 / -- \\
\quad grid & 48 & 39 / 29 / 0.18 & 0 / 0 / -- & 6 / 3 / 0.47 & 3 / 0 / 0.50 \\
\quad context & 48 & 40 / 29 / 0.10 & 1 / 1 / 0.50 & 7 / 2 / 0.10 & 0 / 0 / -- \\
\quad dual & 48 & 39 / 27 / 0.19 & 0 / 2 / -- & 9 / 3 / 0.10 & 0 / 0 / -- \\
\quad ZoomClick-noPZ (public) & 48 & 41 / 30 / 0.18 & 0 / 0 / -- & 6 / 2 / 0.10 & 1 / 0 / 0.50 \\
\addlinespace
\bottomrule
\end{tabular}\end{adjustbox}

\end{table}

\paragraph{Policies.}
Uniform fresh, replay-PPI, generic PPI and RCR all receive the same budget $B$ of fresh target executions; replay-PPI and RCR additionally spend one replay request per maintenance task, which is charged. Best-single picks, from the same uniformly sampled fresh executions, the single skill (or direct) with the highest estimated accuracy and runs it for every task. Costs count every request of every policy, including the direct requests that all policies share; cost curves plot the deployment score against the mean maintenance requests per skill.

\paragraph{Decision outcomes.}
Table~\ref{tab:decisions} lists, per skill, the old (source) model's own estimate on $\mathcal{C}$, from which the carried-over decision is taken, the full-recertification and replay-only estimates of $\tau$ on $\mathcal{C}$, the resulting no-budget decisions, and the fraction of maintenance histories in which each budgeted policy decided $\mathrm{KEEP}$ at the primary budget, next to the sealed-test oracle decision. On the shared pool $\mathcal{C}$ the old and new model's estimates differ in sign for \cnt{summary.uiv.pool-sign-flips} of \cnt{summary.uiv.pool-sign-n} non-zero UI-Vision pairs (\cnt{summary.sspro.pool-sign-flips} of \cnt{summary.sspro.pool-sign-n} on ScreenSpot-Pro), and \cnt{summary.uiv.pool-decision-changes} of \cnt{summary.uiv.pool-decision-n} UI-Vision maintenance decisions changed, each on a new-model estimate of at most \val{summary.uiv.pool-change-max-abs-tau-pt} points (one task); \cnt{summary.uiv.pool-changes-matching-T-oracle} match the sealed oracle (one at the tie) and the rest move away from it, and no sign change holds on $\mathcal{T}$. Including the public ZoomClick-noPZ skill, the counts are \cnt{summary.uiv.pool-sign-flips-with-public} of \cnt{summary.uiv.pool-sign-n-with-public} sign differences (each at a new-model estimate of at most \val{summary.uiv.pool-sign-flip-max-abs-tau-pt-with-public} points; \cnt{summary.uiv.pool-sign-flips-held-on-T-with-public} holding on $\mathcal{T}$) and \cnt{summary.uiv.pool-decision-changes-with-public} of \cnt{summary.uiv.pool-decision-n-with-public} decision changes; the dual skill's catalogue changes pass through an exact-zero estimate. The carried-over decision equals the new model's sealed oracle for \cnt{summary.uiv.p1.old-decision-eq-T-oracle} of \cnt{summary.uiv.p1.old-decision-eq-T-oracle-n} UI-Vision catalogue skills under haiku$\to$sonnet and \cnt{summary.uiv.p2.old-decision-eq-T-oracle} of \cnt{summary.uiv.p2.old-decision-eq-T-oracle-n} under sonnet$\to$opus.

\begin{table}[!htbp]\centering\footnotesize
\caption{Maintenance decisions on $\mathcal{C}$ (frozen before any sealed-test request) and the sealed-test oracle decision (KEEP iff $\tau\ge0$ on $\mathcal{T}$). $\hat\tau$ on $\mathcal{C}$ of the old (source) model's own workflow, of the target's workflow (full recertification) and of replay.}
\label{tab:decisions}
{\setlength{\tabcolsep}{3pt}
\begin{adjustbox}{max width=\linewidth}\begin{tabular}{l rrr ccc rrrr c}
\toprule
 & \multicolumn{3}{c}{$\hat\tau$ on $\mathcal{C}$ (pt)} & \multicolumn{3}{c}{no-budget decision} & \multicolumn{4}{c}{KEEP rate (\%) at 25\%} & on $\mathcal{T}$ \\
Skill & old model & full & replay & old & replay & full & uniform fresh & generic PPI & replay-PPI & RCR & oracle ($\tau\ge0$) \\
\midrule
\multicolumn{12}{l}{\emph{UI-Vision, haiku$\to$sonnet}} \\
\quad center & $-$7.3 & +1.0 & $-$6.2 & SUSP & SUSP & KEEP & 70 & 65 & 50 & 60 & SUSP. \\
\quad grid & $-$14.6 & $-$3.1 & $-$15.6 & SUSP & SUSP & SUSP & 25 & 30 & 40 & 30 & SUSP. \\
\quad context & $-$7.3 & $-$2.1 & $-$1.0 & SUSP & SUSP & SUSP & 45 & 55 & 25 & 10 & SUSP. \\
\quad dual & $-$8.3 & 0.0 & $-$7.3 & SUSP & SUSP & KEEP & 60 & 80 & 50 & 55 & SUSP. \\
\quad ZoomClick-noPZ (public) & $-$3.1 & $-$2.1 & $-$2.1 & SUSP & SUSP & SUSP & 40 & 70 & 35 & 15 & SUSP. \\
\addlinespace
\multicolumn{12}{l}{\emph{UI-Vision, sonnet$\to$opus}} \\
\quad center & +1.0 & +2.1 & +1.0 & KEEP & KEEP & KEEP & 75 & 60 & 70 & 55 & SUSP. \\
\quad grid & $-$3.1 & +1.0 & $-$6.2 & SUSP & SUSP & KEEP & 80 & 55 & 45 & 45 & SUSP. \\
\quad context & $-$2.1 & +1.0 & +2.1 & SUSP & KEEP & KEEP & 60 & 45 & 40 & 55 & KEEP \\
\quad dual & 0.0 & $-$1.0 & $-$2.1 & KEEP & SUSP & SUSP & 80 & 60 & 45 & 40 & SUSP. \\
\quad ZoomClick-noPZ (public) & $-$2.1 & +1.0 & $-$3.1 & SUSP & SUSP & KEEP & 75 & 55 & 55 & 50 & SUSP. \\
\addlinespace
\multicolumn{12}{l}{\emph{ScreenSpot-Pro, haiku$\to$sonnet}} \\
\quad center & +14.6 & +16.7 & +12.5 & KEEP & KEEP & KEEP & 100 & 95 & 100 & 100 & KEEP \\
\quad grid & +20.8 & +14.6 & +12.5 & KEEP & KEEP & KEEP & 90 & 95 & 95 & 100 & KEEP \\
\quad context & +16.7 & +10.4 & +12.5 & KEEP & KEEP & KEEP & 85 & 95 & 100 & 100 & KEEP \\
\quad dual & +12.5 & +12.5 & +6.2 & KEEP & KEEP & KEEP & 95 & 90 & 75 & 95 & KEEP \\
\quad ZoomClick-noPZ (public) & +25.0 & +20.8 & +16.7 & KEEP & KEEP & KEEP & 100 & 100 & 100 & 100 & KEEP \\
\addlinespace
\multicolumn{12}{l}{\emph{ScreenSpot-Pro, sonnet$\to$opus}} \\
\quad center & +16.7 & +10.4 & +8.3 & KEEP & KEEP & KEEP & 100 & 100 & 100 & 100 & KEEP \\
\quad grid & +14.6 & +10.4 & +6.2 & KEEP & KEEP & KEEP & 90 & 85 & 85 & 90 & KEEP \\
\quad context & +10.4 & +10.4 & +10.4 & KEEP & KEEP & KEEP & 95 & 100 & 100 & 100 & KEEP \\
\quad dual & +12.5 & +6.2 & +4.2 & KEEP & KEEP & KEEP & 75 & 60 & 60 & 80 & KEEP \\
\quad ZoomClick-noPZ (public) & +20.8 & +16.7 & +10.4 & KEEP & KEEP & KEEP & 95 & 100 & 100 & 100 & KEEP \\
\addlinespace
\bottomrule
\end{tabular}\end{adjustbox}
}
\end{table}

\section{Additional results, costs and failure cases}
\label{sup:results}
\begin{figure}[!htbp]
\centering
\includegraphics[width=\linewidth]{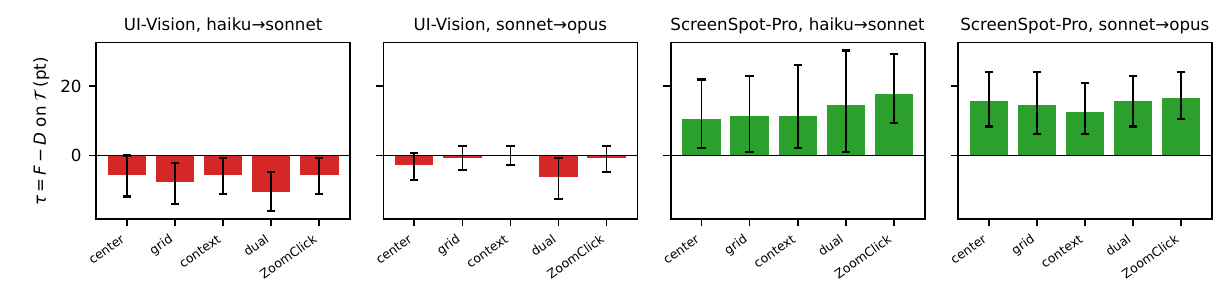}
\caption{Per-skill gain $\hat\tau$ of the target model's own workflow over its direct click on the sealed applications $\mathcal{T}$. Bars: mean gain in points; whiskers: app-level paired-bootstrap 95\% intervals. ZoomClick is the ZoomClick-noPZ adaptation.}
\label{fig:skills}
\end{figure}

\paragraph{Per-skill effects on the sealed test.} Figure~\ref{fig:skills} and Table~\ref{tab:execution} give the sealed-test accuracy of every skill on each target model, its gain over the same model's direct click with an application-level paired-bootstrap interval, the mean number of requests per task, and the direct and direct-self-consistency (medoid of three runs) baselines.

\begin{table}[!htbp]\centering\footnotesize
\caption{Execution layer on the sealed applications $\mathcal{T}$: target-model workflow accuracy, gain $\tau$ over direct (app-level 95\% CI), mean requests per task.}
\label{tab:execution}
\begin{adjustbox}{max width=\linewidth}\begin{tabular}{l r r c r r}
\toprule
Skill (target model) & acc.\ (\%) & $\tau$ (pt) & 95\% CI (app-level) & req./task & $n$ \\
\midrule
\multicolumn{6}{l}{\emph{UI-Vision, haiku$\to$sonnet}: direct 83.3, direct$\times$3 medoid 81.2; $|\mathcal{T}|$ = 144 tasks / 24 apps} \\
\quad center & 77.8 & $-$5.6 & [$-$11.8, 0.0] & 2.01 & 144 \\
\quad grid & 75.7 & $-$7.6 & [$-$13.9, $-$2.1] & 3.65 & 144 \\
\quad context & 77.8 & $-$5.6 & [$-$11.1, $-$0.7] & 2.01 & 144 \\
\quad dual & 72.9 & $-$10.4 & [$-$16.0, $-$4.9] & 3.02 & 144 \\
\quad ZoomClick-noPZ (public) & 77.8 & $-$5.6 & [$-$11.1, $-$0.7] & 3.00 & 144 \\
\addlinespace
\multicolumn{6}{l}{\emph{UI-Vision, sonnet$\to$opus}: direct 84.0, direct$\times$3 medoid 84.7; $|\mathcal{T}|$ = 144 tasks / 24 apps} \\
\quad center & 81.2 & $-$2.8 & [$-$6.9, +0.7] & 2.00 & 144 \\
\quad grid & 83.3 & $-$0.7 & [$-$4.2, +2.8] & 3.58 & 144 \\
\quad context & 84.0 & 0.0 & [$-$2.8, +2.8] & 2.00 & 144 \\
\quad dual & 77.8 & $-$6.2 & [$-$12.5, $-$0.7] & 3.00 & 144 \\
\quad ZoomClick-noPZ (public) & 83.3 & $-$0.7 & [$-$4.9, +2.8] & 3.00 & 144 \\
\addlinespace
\multicolumn{6}{l}{\emph{ScreenSpot-Pro, haiku$\to$sonnet}: direct 74.0, direct$\times$3 medoid 72.9; $|\mathcal{T}|$ = 96 tasks / 12 apps} \\
\quad center & 84.4 & +10.4 & [+2.1, +21.9] & 2.00 & 96 \\
\quad grid & 85.4 & +11.5 & [+1.0, +22.9] & 3.84 & 96 \\
\quad context & 85.4 & +11.5 & [+2.1, +26.0] & 2.00 & 96 \\
\quad dual & 88.5 & +14.6 & [+1.0, +30.2] & 3.02 & 96 \\
\quad ZoomClick-noPZ (public) & 91.7 & +17.7 & [+9.4, +29.2] & 2.98 & 96 \\
\addlinespace
\multicolumn{6}{l}{\emph{ScreenSpot-Pro, sonnet$\to$opus}: direct 76.0, direct$\times$3 medoid 81.2; $|\mathcal{T}|$ = 96 tasks / 12 apps} \\
\quad center & 91.7 & +15.6 & [+8.3, +24.0] & 2.00 & 96 \\
\quad grid & 90.6 & +14.6 & [+6.2, +24.0] & 3.68 & 96 \\
\quad context & 88.5 & +12.5 & [+6.2, +20.8] & 2.00 & 96 \\
\quad dual & 91.7 & +15.6 & [+8.3, +22.9] & 3.00 & 96 \\
\quad ZoomClick-noPZ (public) & 92.7 & +16.7 & [+10.4, +24.0] & 3.00 & 96 \\
\addlinespace
\bottomrule
\end{tabular}\end{adjustbox}

\end{table}

\begin{figure}[!htbp]
\centering
\includegraphics[width=\linewidth]{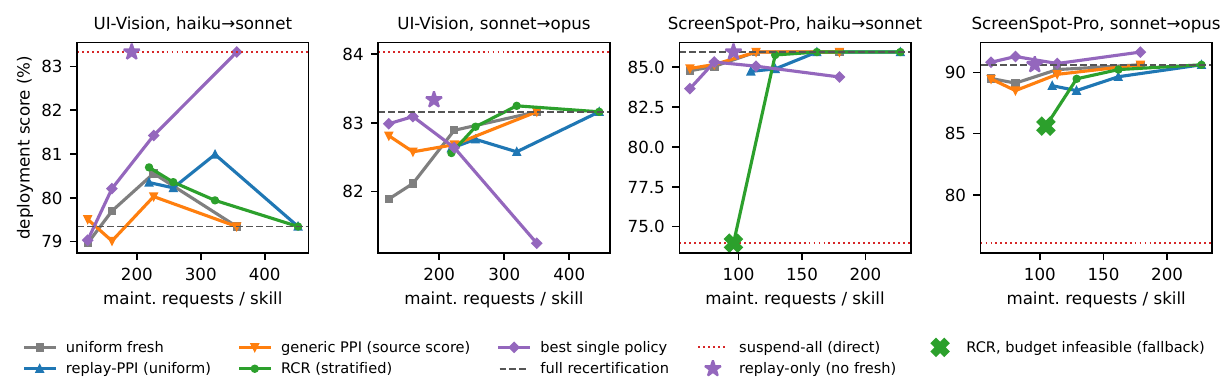}
\caption{Catalogue deployment score on $\mathcal{T}$ against mean maintenance requests per skill on $\mathcal{C}$ for the budgeted policies at 10/25/50/100\% budgets (mean over 20 histories). Lines: full recertification and suspend-all; star: replay-only at its cost. Crosses: RCR budget points at which the stratified allocation was infeasible in some histories (fallback to direct), plotted at the requests actually spent.}
\label{fig:cost}
\end{figure}

\paragraph{Cost curves.} Figure~\ref{fig:cost} traces the catalogue deployment score against maintenance requests for every budgeted policy and budget.

\paragraph{Preregistered comparisons.} Table~\ref{tab:comparisons} lists all comparisons at the primary budget. Holm-adjusted $p$-values are over the family of \cnt{holm.family-size} one-sided superiority tests (two comparisons, two domains, two model changes), the more conservative of the two readings the frozen plan allows; this choice and the one-sided bootstrap $p$-value were added to the analysis code after the evaluation had printed unadjusted lower bounds and before any result was written up (decision D-018 of the project decision log). The non-inferiority test uses its preregistered one-sided 95\% lower bound without multiplicity correction. Leave-one-sealed-application-out: across the \cnt{summary.all.loo-comparisons} primary comparisons, dropping any single sealed application leaves the mean difference non-negative in every case (\cnt{summary.all.loo-n-negative} negative; smallest value \ppt{summary.all.loo-min} points). These intervals and tests are conditional on the realised maintenance pool; Table~\ref{tab:estimator} and Sec.~\ref{sup:robust} report the pool-level sensitivity. Applying Holm per setting (families of two) gives the same rejections with the 20 frozen histories (\cnt{rev.pol.holm.f20.per-setting.n-rejected}) and \cnt{rev.pol.holm.n.per-setting.n-rejected} with \cnt{rev.pol.n-hist} histories (Sec.~\ref{sup:revision}). The bootstrap seeds were made content-derived after the second internal review (D-021), so that identical comparisons receive identical intervals; this moved \cnt{holm.n-marginal} adjusted $p$-values from just below to just at the edge of $\alpha$ without changing any rejection, and these are the marginal ones; with $B=\cnt{rev.pol.boot-B-hi}$ they stay rejected under three seed schemes, so their fragility comes from the 20-history average, not from bootstrap error (Sec.~\ref{sup:revision}). At the same review the maintenance-cost axis was restricted to the catalogue skills to match the catalogue score (D-021).

\begin{table}[!htbp]\centering\footnotesize
\caption{Comparisons at the primary budget and equal fresh budget, not cost-matched (see Table~\ref{tab:equalcost} for the equal-cost comparison); catalogue score difference in points; app-level paired bootstrap. $^{*}$: rejected at $\alpha=0.05$ after Holm over the \cnt{holm.family-size} preregistered superiority tests, with the 20 frozen histories; with \cnt{rev.pol.n-hist} histories \cnt{rev.pol.holm.n-rejected-both} of these rejections remain (Table~\ref{tab:revholm}).}
\label{tab:comparisons}
\begin{adjustbox}{max width=\linewidth}\begin{tabular}{l r c r r r}
\toprule
Comparison (catalogue score, pt) & mean & 95\% CI & one-sided 95\% LB & Holm $p$ & apps \\
\midrule
\multicolumn{6}{l}{\emph{UI-Vision, haiku$\to$sonnet} (24 apps)} \\
\quad RCR $-$ replay-PPI & +0.1 & [$-$0.1, +0.4] & $-$0.1 & 0.165 & 24 \\
\quad RCR $-$ generic PPI & +1.3 & [+0.7, +2.1] & +0.8 & 0.002$^{*}$ & 24 \\
\quad RCR $-$ uniform fresh & +0.7 & [+0.2, +1.2] & +0.3 & -- & 24 \\
\quad RCR $-$ best-single & +0.1 & [$-$1.0, +1.4] & $-$0.8 & -- & 24 \\
\quad RCR $-$ suspend-all & $-$3.0 & [$-$4.7, $-$1.5] & $-$4.4 & -- & 24 \\
\quad RCR $-$ full recert.\ (NI margin $-$1 pt) & +1.0 & [+0.2, +1.8] & +0.3 & -- & 24 \\
\addlinespace
\multicolumn{6}{l}{\emph{UI-Vision, sonnet$\to$opus} (24 apps)} \\
\quad RCR $-$ replay-PPI & +0.2 & [+0.02, +0.4] & +0.04 & 0.049$^{*}$ & 24 \\
\quad RCR $-$ generic PPI & +0.4 & [+0.03, +0.7] & +0.1 & 0.049$^{*}$ & 24 \\
\quad RCR $-$ uniform fresh & +0.8 & [$-$0.1, +1.8] & +0.1 & -- & 24 \\
\quad RCR $-$ best-single & $-$0.1 & [$-$0.9, +0.6] & $-$0.8 & -- & 24 \\
\quad RCR $-$ suspend-all & $-$1.1 & [$-$2.7, +0.3] & $-$2.4 & -- & 24 \\
\quad RCR $-$ full recert.\ (NI margin $-$1 pt) & $-$0.2 & [$-$0.9, +0.6] & $-$0.8 & -- & 24 \\
\addlinespace
\multicolumn{6}{l}{\emph{ScreenSpot-Pro, haiku$\to$sonnet} (12 apps)} \\
\quad RCR $-$ replay-PPI & +0.9 & [+0.1, +1.8] & +0.2 & 0.049$^{*}$ & 12 \\
\quad RCR $-$ generic PPI & +0.6 & [+0.1, +1.3] & +0.2 & 0.021$^{*}$ & 12 \\
\quad RCR $-$ uniform fresh & +0.7 & [+0.1, +1.5] & +0.2 & -- & 12 \\
\quad RCR $-$ best-single & +0.4 & [$-$1.8, +3.0] & $-$1.5 & -- & 12 \\
\quad RCR $-$ suspend-all & +11.8 & [+1.8, +23.8] & +3.1 & -- & 12 \\
\quad RCR $-$ full recert.\ (NI margin $-$1 pt) & $-$0.2 & [$-$0.4, $-$0.01] & $-$0.3 & -- & 12 \\
\addlinespace
\multicolumn{6}{l}{\emph{ScreenSpot-Pro, sonnet$\to$opus} (12 apps)} \\
\quad RCR $-$ replay-PPI & +1.0 & [+0.5, +1.4] & +0.6 & 0.002$^{*}$ & 12 \\
\quad RCR $-$ generic PPI & +1.0 & [+0.5, +1.4] & +0.6 & 0.002$^{*}$ & 12 \\
\quad RCR $-$ uniform fresh & +0.4 & [+0.2, +0.5] & +0.2 & -- & 12 \\
\quad RCR $-$ best-single & $-$1.8 & [$-$2.8, $-$0.9] & $-$2.7 & -- & 12 \\
\quad RCR $-$ suspend-all & +13.4 & [+7.2, +20.9] & +8.1 & -- & 12 \\
\quad RCR $-$ full recert.\ (NI margin $-$1 pt) & $-$1.1 & [$-$1.7, $-$0.7] & $-$1.6 & -- & 12 \\
\addlinespace
\bottomrule
\end{tabular}\end{adjustbox}

\end{table}

\begin{table}[!htbp]\centering\footnotesize
\begin{adjustbox}{max width=\linewidth}\begin{tabular}{l cccc}
\toprule
 & \multicolumn{4}{c}{$N$ / share of tasks / mean residual $F-R$ (pt), catalogue skills pooled} \\
Setting & agree/stable & agree/drift & disagree & parse fail \\
\midrule
UI-Vision, haiku$\to$sonnet, $\mathcal{C}$ & 271 / 71\% / +1.8 & 36 / 9\% / $-$11.1 & 70 / 18\% / +27.1 & 7 / 2\% / +71.4$^{\dagger}$ \\
UI-Vision, sonnet$\to$opus, $\mathcal{C}$ & 334 / 87\% / +0.6 & 9 / 2\% / +33.3$^{\dagger}$ & 39 / 10\% / +2.6 & 2 / 1\% / +100.0$^{\dagger}$ \\
UI-Vision, sonnet$\to$opus, $\mathcal{T}$ & 512 / 89\% / $-$0.8 & 11 / 2\% / 0.0 & 47 / 8\% / +19.1 & 6 / 1\% / +50.0$^{\dagger}$ \\
ScreenSpot-Pro, haiku$\to$sonnet, $\mathcal{C}$ & 146 / 76\% / +1.4 & 7 / 4\% / 0.0$^{\dagger}$ & 33 / 17\% / +3.0 & 6 / 3\% / +33.3$^{\dagger}$ \\
ScreenSpot-Pro, sonnet$\to$opus, $\mathcal{C}$ & 160 / 83\% / $-$1.2 & 1 / 1\% / 0.0$^{\dagger}$ & 28 / 15\% / +10.7 & 3 / 2\% / +100.0$^{\dagger}$ \\
ScreenSpot-Pro, sonnet$\to$opus, $\mathcal{T}$ & 334 / 87\% / +0.9 & 6 / 2\% / 0.0$^{\dagger}$ & 41 / 11\% / +9.8 & 3 / 1\% / +100.0$^{\dagger}$ \\
\bottomrule
\end{tabular}\end{adjustbox}

\caption{Replay residual by frozen stratum ($N$ / share of tasks / mean $F-R$ in points), catalogue skills pooled, on the maintenance pool $\mathcal{C}$ and the sealed test $\mathcal{T}$. $^\dagger$: fewer than ten tasks.}
\label{tab:bias}
\end{table}

\paragraph{Replay residual by stratum.} Table~\ref{tab:bias} gives the residual per frozen stratum. The gap between the disagreement and agree/stable residuals, with app-level cluster-bootstrap 95\% intervals (exploratory), is on the maintenance pools \ppt{expl.uiv.p1.e3.C.gap} [\ppt{expl.uiv.p1.e3.C.ci95.0}, \ppt{expl.uiv.p1.e3.C.ci95.1}] (UI-Vision haiku$\to$sonnet), \ppt{expl.uiv.p2.e3.C.gap} [\ppt{expl.uiv.p2.e3.C.ci95.0}, \ppt{expl.uiv.p2.e3.C.ci95.1}] (UI-Vision sonnet$\to$opus), \ppt{expl.sspro.p1.e3.C.gap} [\ppt{expl.sspro.p1.e3.C.ci95.0}, \ppt{expl.sspro.p1.e3.C.ci95.1}] (ScreenSpot-Pro haiku$\to$sonnet) and \ppt{expl.sspro.p2.e3.C.gap} [\ppt{expl.sspro.p2.e3.C.ci95.0}, \ppt{expl.sspro.p2.e3.C.ci95.1}] (ScreenSpot-Pro sonnet$\to$opus); the last upper bound reflects very small cells.

\paragraph{Reverse-direction readers.} Table~\ref{tab:reverse} gives, for the preregistered subset of sealed tasks, the accuracy of the source model on its own direct click and when it reads the target model's packet.

\begin{table}[!htbp]\centering\footnotesize
{
\begin{adjustbox}{max width=\linewidth}\begin{tabular}{l r r c r}
\toprule
Skill & reader direct (\%) & reader on packet (\%) & gain (pt) [95\% CI] & $n$ \\
\midrule
\multicolumn{5}{l}{\emph{UI-Vision: haiku reads sonnet's packets}} \\
\quad center & 63.3 & 65.0 & +1.7 [$-$3.3, +7.5] & 120 \\
\quad grid & 63.3 & 65.8 & +2.5 [$-$4.2, +9.2] & 120 \\
\quad context & 63.3 & 63.3 & 0.0 [$-$5.8, +5.8] & 120 \\
\quad dual & 63.3 & 67.5 & +4.2 [$-$2.5, +10.0] & 120 \\
\quad ZoomClick-noPZ (public) & 63.3 & 69.2 & +5.8 [0.0, +11.7] & 120 \\
\addlinespace
\multicolumn{5}{l}{\emph{UI-Vision: sonnet reads opus's packets}} \\
\quad center & 82.5 & 76.7 & $-$5.8 [$-$11.7, $-$0.8] & 120 \\
\quad grid & 82.5 & 79.2 & $-$3.3 [$-$9.2, +1.7] & 120 \\
\quad context & 82.5 & 76.7 & $-$5.8 [$-$11.7, $-$0.8] & 120 \\
\quad dual & 82.5 & 78.3 & $-$4.2 [$-$10.0, +0.8] & 120 \\
\quad ZoomClick-noPZ (public) & 82.5 & 77.5 & $-$5.0 [$-$10.0, $-$0.8] & 120 \\
\addlinespace
\multicolumn{5}{l}{\emph{ScreenSpot-Pro: haiku reads sonnet's packets}} \\
\quad center & 47.9 & 68.8 & +20.8 [+8.3, +33.3] & 48 \\
\quad grid & 47.9 & 77.1 & +29.2 [+14.6, +45.8] & 48 \\
\quad context & 47.9 & 77.1 & +29.2 [+12.5, +45.8] & 48 \\
\quad dual & 47.9 & 77.1 & +29.2 [+8.3, +52.1] & 48 \\
\quad ZoomClick-noPZ (public) & 47.9 & 85.4 & +37.5 [+20.8, +54.2] & 48 \\
\addlinespace
\multicolumn{5}{l}{\emph{ScreenSpot-Pro: sonnet reads opus's packets}} \\
\quad center & 75.0 & 85.4 & +10.4 [$-$6.2, +29.2] & 48 \\
\quad grid & 75.0 & 87.5 & +12.5 [$-$4.2, +31.2] & 48 \\
\quad context & 75.0 & 87.5 & +12.5 [$-$4.2, +31.2] & 48 \\
\quad dual & 75.0 & 93.8 & +18.8 [+6.2, +33.3] & 48 \\
\quad ZoomClick-noPZ (public) & 75.0 & 91.7 & +16.7 [+4.2, +29.2] & 48 \\
\addlinespace
\bottomrule
\end{tabular}\end{adjustbox}
}
\caption{Reverse-direction readers on the sealed-test subset: the source model reads the target model's packet.}
\label{tab:reverse}
\end{table}

\paragraph{Estimation on the maintenance pool and robustness to the pool.}
\label{sup:robust}
Table~\ref{tab:estimator} gives, for every budgeted policy, the error of $\hat\tau$ against the full-pool estimate on $\mathcal{C}$ and the agreement of its decision with full recertification, and the leave-one-$\mathcal{C}$-application-out counts. A two-stage bootstrap that resamples maintenance and sealed applications and reruns the decisions (exploratory; \cnt{expl.uiv.p1.e6.reps} replicates of \cnt{expl.uiv.p1.e6.histories} histories) gives RCR minus replay-PPI between \ppt{summary.e6.replay-ppi-uniform.min} and \ppt{summary.e6.replay-ppi-uniform.max} points and RCR minus generic PPI between \ppt{summary.e6.generic-ppi-source.min} and \ppt{summary.e6.generic-ppi-source.max} points across the four settings, and RCR minus replay-only between \ppt{summary.e6.replay-only.min} and \ppt{summary.e6.replay-only.max} points, with 95\% intervals that include zero in \cnt{summary.e6.n-ci-includes-0} of \cnt{summary.e6.n} comparisons (against uniform fresh, replay-PPI, generic PPI and replay-only); both the equal-budget advantages of RCR and its deficit against replay-only are therefore not shown beyond the realised pool. In estimation, the uncorrected replay estimate has a lower error than RCR in \cnt{summary.expl.replay-only-lower-rmse-than-rcr} of \cnt{summary.expl.replay-only-rmse-n} settings (Table~\ref{tab:estimator}).

\paragraph{Within-reader contrast (exploratory).} Table~\ref{tab:within} compares, on the same sealed tasks, sonnet reading opus's packets with sonnet's own workflow.

\begin{table}[!htbp]\centering\footnotesize
{
\begin{adjustbox}{max width=\linewidth}\begin{tabular}{l r r c}
\toprule
Skill & reads opus's packet (\%) & own workflow (\%) & difference (pt) [95\% CI] \\
\midrule
\multicolumn{4}{l}{\emph{UI-Vision: sonnet reads opus's packet vs.\ sonnet's own workflow (same tasks)}} \\
\quad center & 76.7 & 77.5 & $-$0.8 [$-$5.0, +3.3] \\
\quad grid & 79.2 & 75.0 & +4.2 [$-$1.7, +10.0] \\
\quad context & 76.7 & 76.7 & 0.0 [$-$4.2, +4.2] \\
\quad dual & 78.3 & 72.5 & +5.8 [+0.8, +11.7] \\
\quad ZoomClick-noPZ (public) & 77.5 & 78.3 & $-$0.8 [$-$5.8, +4.2] \\
\addlinespace
\multicolumn{4}{l}{\emph{ScreenSpot-Pro: sonnet reads opus's packet vs.\ sonnet's own workflow (same tasks)}} \\
\quad center & 85.4 & 81.2 & +4.2 [$-$6.2, +14.6] \\
\quad grid & 87.5 & 81.2 & +6.2 [0.0, +12.5] \\
\quad context & 87.5 & 81.2 & +6.2 [0.0, +12.5] \\
\quad dual & 93.8 & 83.3 & +10.4 [+4.2, +16.7] \\
\quad ZoomClick-noPZ (public) & 91.7 & 91.7 & 0.0 [0.0, 0.0] \\
\addlinespace
\bottomrule
\end{tabular}\end{adjustbox}
}
\caption{Exploratory within-reader contrast on the reverse-reader subset: sonnet reading opus's packet vs.\ sonnet's own workflow on the same tasks.}
\label{tab:within}
\end{table}

\begin{table}[!htbp]\centering\footnotesize
\caption{Exploratory: estimation quality of the budgeted policies on $\mathcal{C}$ (RMSE of $\hat\tau$ against the full-pool $\hat\tau$, in points, over all five skills including the public one and 20 histories / agreement with the full-recertification decision), the same at RCR's request cost (linear interpolation between budget points, conservative for the competitors because error is convex in budget), the two estimates that need no fresh execution, and leave-one-$\mathcal{C}$-application-out counts of settings where RCR scores below each competitor on $\mathcal{T}$.}
\label{tab:estimator}
{\setlength{\tabcolsep}{3pt}
\begin{adjustbox}{max width=\linewidth}\begin{tabular}{l cccc}
\toprule
Setting, budget & uniform fresh & generic PPI & replay-PPI & RCR \\
 & \multicolumn{4}{c}{RMSE (pt) / agreement (\%)} \\
\midrule
UI-Vision, haiku$\to$sonnet, 10\% & 10.3 / 56 & 12.3 / 47 & 9.8 / 70 & 6.7 / 65 \\
UI-Vision, haiku$\to$sonnet, 25\% & 5.7 / 64 & 8.0 / 58 & 6.3 / 60 & 4.8 / 72 \\
UI-Vision, haiku$\to$sonnet, 50\% & 3.8 / 68 & 4.5 / 61 & 3.3 / 70 & 2.6 / 79 \\
\multicolumn{5}{l}{\quad no fresh execution: replay-only 7.3 / 60; old (source) estimate 7.7 / 60} \\
\quad at RCR's request cost (257, interpolated) & 2.9 / 76 & 3.4 / 70 & 6.3 / 60 & 4.8 / 72 \\
\multicolumn{5}{l}{\quad leave-one-$\mathcal{C}$-app-out, settings where RCR is below uniform / generic / replay-PPI / replay-only: 0/12 / 2/12 / 4/12 / 12/12} \\
\addlinespace
UI-Vision, sonnet$\to$opus, 10\% & 9.6 / 68 & 11.1 / 55 & 10.8 / 58 & 9.7 / 48 \\
UI-Vision, sonnet$\to$opus, 25\% & 5.7 / 62 & 6.5 / 51 & 6.2 / 53 & 4.6 / 53 \\
UI-Vision, sonnet$\to$opus, 50\% & 2.8 / 76 & 3.2 / 71 & 3.0 / 65 & 2.8 / 55 \\
\multicolumn{5}{l}{\quad no fresh execution: replay-only 3.8 / 60; old (source) estimate 2.8 / 20} \\
\quad at RCR's request cost (256, interpolated) & 2.1 / 82 & 2.4 / 78 & 6.2 / 53 & 4.6 / 53 \\
\multicolumn{5}{l}{\quad leave-one-$\mathcal{C}$-app-out, settings where RCR is below uniform / generic / replay-PPI / replay-only: 1/12 / 7/12 / 3/12 / 10/12} \\
\addlinespace
ScreenSpot-Pro, haiku$\to$sonnet, 10\% & 20.9 / 91 & 13.8 / 93 & 12.0 / 92 & infeasible \\
ScreenSpot-Pro, haiku$\to$sonnet, 25\% & 12.2 / 94 & 8.2 / 95 & 6.6 / 94 & 7.7 / 99 \\
ScreenSpot-Pro, haiku$\to$sonnet, 50\% & 6.5 / 100 & 4.4 / 100 & 3.8 / 100 & 3.9 / 100 \\
\multicolumn{5}{l}{\quad no fresh execution: replay-only 4.1 / 100; old (source) estimate 4.5 / 100} \\
\quad at RCR's request cost (129, interpolated) & 5.0 / 100 & 3.3 / 100 & 6.6 / 94 & 7.7 / 99 \\
\multicolumn{5}{l}{\quad leave-one-$\mathcal{C}$-app-out, settings where RCR is below uniform / generic / replay-PPI / replay-only: 2/6 / 3/6 / 2/6 / 5/6} \\
\addlinespace
ScreenSpot-Pro, sonnet$\to$opus, 10\% & 15.5 / 93 & 13.1 / 92 & 12.4 / 91 & 11.1 / 91 \\
ScreenSpot-Pro, sonnet$\to$opus, 25\% & 11.4 / 91 & 8.7 / 89 & 8.0 / 89 & 5.2 / 94 \\
ScreenSpot-Pro, sonnet$\to$opus, 50\% & 5.1 / 98 & 4.9 / 96 & 4.9 / 95 & 3.5 / 98 \\
\multicolumn{5}{l}{\quad no fresh execution: replay-only 3.6 / 100; old (source) estimate 4.8 / 100} \\
\quad at RCR's request cost (129, interpolated) & 3.9 / 98 & 3.7 / 97 & 8.0 / 89 & 5.2 / 94 \\
\multicolumn{5}{l}{\quad leave-one-$\mathcal{C}$-app-out, settings where RCR is below uniform / generic / replay-PPI / replay-only: 5/6 / 2/6 / 5/6 / 6/6} \\
\addlinespace
\bottomrule
\end{tabular}\end{adjustbox}
}
\end{table}

\paragraph{Visual diagnostic by packet type.} Table~\ref{tab:visiondetail} splits the missed-target tasks of Table~\ref{tab:vision} by packet type and gives, for the label-free flag, where flagged tasks fall among the frozen strata and the permutation $p$ of the refinement test with flags shuffled across (task, skill) rows and, as reported in the main paper, only within each application (2000 permutations each; exploratory, unadjusted).

\begin{table}[!htbp]\centering\footnotesize
\caption{Exploratory (rows without a completed source packet excluded, as in Table~\ref{tab:vision}): missed-target tasks by packet type (single zoomed view vs.\ views with a global thumbnail; $N$ and replay/workflow accuracy) and the label-free flag (flagged rows, share outside agree/stable and inside the disagreement cell, permutation $p$).}
\label{tab:visiondetail}
\begin{adjustbox}{max width=\linewidth}\begin{tabular}{l c c r r r r r}
\toprule
 & \multicolumn{2}{c}{missed targets: $N$ / $R$/$F$ (\%)} & \multicolumn{3}{c}{label-free flag} & \multicolumn{2}{c}{permutation $p$} \\
\cmidrule(lr){2-3}\cmidrule(lr){4-6}\cmidrule(lr){7-8}
Setting & single view & with thumbnail & $N$ & out.\ agr./st.\ \% & in disagr.\ \% & row-wise & within app \\
\midrule
UI-Vision, haiku$\to$sonnet, $\mathcal{C}$ & 35 / 0.0/45.7 & 32 / 15.6/43.8 & 45 & 96 & 76 & 0.003 & 0.030 \\
UI-Vision, sonnet$\to$opus, $\mathcal{C}$ & 17 / 0.0/35.3 & 18 / 16.7/27.8 & 20 & 100 & 85 & 0.223 & 0.827 \\
UI-Vision, sonnet$\to$opus, $\mathcal{T}$ & 21 / 4.8/23.8 & 24 / 12.5/20.8 & 19 & 95 & 79 & 0.196 & 0.168 \\
ScreenSpot-Pro, haiku$\to$sonnet, $\mathcal{C}$ & 12 / 0.0/16.7 & 12 / 16.7/25.0 & 10 & 100 & 100 & 1.000 & 1.000 \\
ScreenSpot-Pro, sonnet$\to$opus, $\mathcal{C}$ & 12 / 0.0/8.3 & 15 / 6.7/46.7 & 21 & 100 & 90 & 0.673 & 0.716 \\
ScreenSpot-Pro, sonnet$\to$opus, $\mathcal{T}$ & 8 / 0.0/37.5 & 6 / 0.0/16.7 & 17 & 100 & 100 & 1.000 & 1.000 \\
\bottomrule
\end{tabular}\end{adjustbox}

\end{table}

\paragraph{Infrastructure failures and execution incidents.}
\label{sup:infra}
In the two main protocols the harvester recorded \cnt{all.records.status.ok} completed requests, \cnt{all.records.status.harness-failed} infrastructure failures and \cnt{all.records.status.missing-in-journal} requests that were never started because a wave was stopped; of the \cnt{d020.n-total} zero-read answers described below, the \cnt{d020.n-T} reclassified ones are among the failures and the other \cnt{d020.n-C} among the completed requests; the development pilots are counted separately in the research artifact. Infrastructure failures were of three kinds: usage-limit errors of the host and file-system read timeouts during one incident late in the sealed test, when the project directory was evicted to cloud storage under disk pressure and some actors could not read their request or images (decision D-016); and \cnt{d020.n-total} answers returned in one short window by the smallest model without a single file read (D-020). Usage-limit and read-timeout failures were re-queued unchanged and are excluded from all counts and analyses. The zero-read rule (D-020) and its re-runs were adopted after the sealed-test results had been computed and reviewed. Of the zero-read answers, the \cnt{d020.n-T} sealed-test requests (reverse-reader replays and direct clicks of the source model) were reclassified and re-run; the \cnt{d020.n-C} development and maintenance answers feed the frozen decisions and remain in the counts, the decisions and the maintenance-pool rows of Table~\ref{tab:bias} and Table~\ref{tab:vision}. Rerunning the primary-budget decisions with the affected maintenance (task, skill) pairs removed changes the UI-Vision haiku$\to$sonnet catalogue scores by at most \ppt{d020.sensitivity.max-abs-change} points and moves RCR minus replay-PPI from \ppt{d020.sensitivity.rcr-minus-replay-ppi-uniform.base} to \ppt{d020.sensitivity.rcr-minus-replay-ppi-uniform.excluding} points and RCR minus generic PPI from \ppt{d020.sensitivity.rcr-minus-generic-ppi-source.base} to \ppt{d020.sensitivity.rcr-minus-generic-ppi-source.excluding} points; the development answers among them also entered $\sigma_h$; re-estimating $\sigma_h$ without them changes \cnt{rev.misc.zr.sigma-extra-changes} further budgeted decision (Sec.~\ref{sup:revision}). In record files the protocol hash is the SHA-256 of the canonical (key-sorted) protocol JSON, while the decision log records the file hash; both identify the same files. In \cnt{audit.partial-image-reads} completed requests the model read its request but not every listed image; these are model behaviour and are counted. Two tooling safeguards were added during the sealed test without changing any request: a pre-launch check that every file a wave reads is local, and a check that every agent ran on the model alias its request specified (D-017; no mismatch in any wave). Unexpected tool calls by actors were harness-interface calls or shell echoes (of the model's own answer or a fixed no-op string) (\cnt{uiv.records.audit.harness-meta-tool-calls-only} and \cnt{sspro.records.audit.harness-meta-tool-calls-only}), read attempts on non-existent paths (\cnt{uiv.records.audit.file-access-attempts-nonexistent} and \cnt{sspro.records.audit.file-access-attempts-nonexistent}), and shell calls other than an echo (\cnt{uiv.records.audit.bash-non-echo} and \cnt{sspro.records.audit.bash-non-echo}), the last of which opened an allowed view and returned nothing to the model; all are listed in the research artifact with their classification. Decision identifiers (D-xxx) refer to entries of the project decision log.

\paragraph{Reverse readers (descriptive; Tables~\ref{tab:reverse}--\ref{tab:within}).} Against its own direct click, haiku reading sonnet's packets on ScreenSpot-Pro gains \ppt{bias.sspro.p1.reverse-reader-T.center-d1.gain.mean} to \ppt{bias.sspro.p1.reverse-reader-T.zoomclick-d3.gain.mean} points, while sonnet reading opus's packets on UI-Vision changes by \ppt{bias.uiv.p2.reverse-reader-T.center-d1.gain.mean} to \ppt{bias.uiv.p2.reverse-reader-T.grid2-d3.gain.mean}. Against sonnet's own workflow on the same tasks (exploratory), opus's packets score \ppt{summary.e4.sspro.min} to \ppt{summary.e4.sspro.max} points on ScreenSpot-Pro and \ppt{summary.e4.uiv.min} to \ppt{summary.e4.uiv.max} on UI-Vision.

\section{Reproduction}
\label{sup:repro}
The research artifact contains the data census and split code (seeded), the image contract and skill programs, the request scheduler with its request ledger, the harvester that parses harness journals and agent transcripts into per-request records (model identifier, provider request identifiers, token usage, tool-use audit, image byte hashes), the scorer, the frozen analysis plan, the decision and evaluation code, and the figure, table and number exporters. Every model answer is stored, so everything from scoring onwards is regenerated offline by a documented sequence of commands; the decision files must reproduce byte for byte against the hashes in the decision log, and every measured or derived number in this paper is a macro generated from the exported analysis files, and only design constants are typed. Re-issuing requests needs model access: request files are plain JSON (instruction, ordered image paths with displayed sizes, output contract) and can be sent through any API adapter that returns the JSON object; the harness we used is not seedable, so re-execution yields new samples, which is why all results are reported with application-level intervals. The research artifact provides screenshot download scripts and image lists instead of the full screenshot sets; the answer keys are kept in an evaluator-only directory that no request references. Environment: Python~3.12 with Pillow and matplotlib, no GPU; the two main protocols used \cnt{all.records.api-calls} harness API calls.

\section{Revision analyses (post hoc)}
\label{sup:revision}
All analyses in this section were added in a revision, after the sealed results and a review. They use only the recorded outcomes, with no new model request, are produced by \texttt{research/analysis/\allowbreak revision\_policy.py} and \texttt{revision\_desc.py}, and are exploratory; the frozen results above remain the preregistered record. Displayed values are the exact values rounded to the nearest displayed digit with ties to even (checked on \cnt{rev.desc.d8.n-display-equals-half-even} boundary cases).

\paragraph{Replay bias per skill.} Table~\ref{tab:revagree} gives the agree/stable residual per skill with application-level intervals (\cnt{rev.desc.summary.d3.reproduced} of \cnt{rev.desc.summary.d3.reproduce-n} published cells reproduced). No per-skill interval excludes zero, but they reach \ppt{rev.desc.summary.d3.perskill-ci.lo-min} to \ppt{rev.desc.summary.d3.perskill-ci.hi-max} points. The agreement cut-off of the strata is about \val{rev.desc.d4.uiv.all.agree-threshold-over-target-short-side.median} (UI-Vision) and \val{rev.desc.d4.sspro.all.agree-threshold-over-target-short-side.median} (ScreenSpot-Pro) times a target's shorter side; re-stratifying with a cut-off of one target diagonal (which uses the annotated target size and is not available at deployment) gives pooled agree/stable residuals of \ppt{rev.desc.summary.d6.tdiag.pooled-resid.min} to \ppt{rev.desc.summary.d6.tdiag.pooled-resid.max} points (Table~\ref{tab:revrestrat}); a same-box variant uses the answer key and forces $F-R\le0$, so it is reported only for completeness.

\begin{table}[!htbp]\centering\scriptsize
\begin{adjustbox}{max width=\linewidth}\begin{tabular}{l l r r c r c}
\toprule
Setting & Skill & $n$ & $N_{\mathrm{a/s}}$ & share (\%) [95\% CI] & $F-R$ (pt) & 95\% CI \\
\midrule
UI-Vision, haiku$\to$sonnet, $\mathcal{C}$ & center & 96 & 70 & 72.9 [65.6, 80.2] & $+$1.4 & [0.0, $+$4.2] \\
 & grid & 96 & 62 & 64.6 [54.2, 75.0] & $+$3.2 & [$-$3.0, $+$9.3] \\
 & context & 96 & 73 & 76.0 [67.7, 84.4] & $+$1.4 & [0.0, $+$4.1] \\
 & dual & 96 & 66 & 68.8 [62.5, 75.0] & $+$1.5 & [0.0, $+$5.0] \\
 & ZoomClick-noPZ & 96 & 76 & 79.2 [70.8, 86.5] & $-$1.3 & [$-$5.8, $+$2.7] \\
 & pooled (cat.) & 384 & 271 & 70.6 [64.1, 77.6] & $+$1.8 & [$-$0.4, $+$4.2] \\
\addlinespace
UI-Vision, sonnet$\to$opus, $\mathcal{C}$ & center & 96 & 87 & 90.6 [85.4, 95.8] & 0.0 & [$-$3.3, $+$3.6] \\
 & grid & 96 & 77 & 80.2 [71.9, 87.5] & $+$2.6 & [$-$2.6, $+$7.7] \\
 & context & 96 & 84 & 87.5 [81.2, 93.8] & 0.0 & [$-$3.4, $+$3.7] \\
 & dual & 96 & 86 & 89.6 [82.3, 95.8] & 0.0 & [$-$5.7, $+$7.1] \\
 & ZoomClick-noPZ & 96 & 85 & 88.5 [82.3, 94.8] & $+$2.4 & [$-$2.2, $+$7.3] \\
 & pooled (cat.) & 384 & 334 & 87.0 [81.5, 91.9] & $+$0.6 & [$-$3.0, $+$4.3] \\
\addlinespace
UI-Vision, sonnet$\to$opus, $\mathcal{T}$ & center & 144 & 128 & 88.9 [81.2, 95.8] & $-$0.8 & [$-$2.4, 0.0] \\
 & grid & 144 & 124 & 86.1 [78.5, 92.4] & $-$1.6 & [$-$4.2, 0.0] \\
 & context & 144 & 131 & 91.0 [85.4, 95.8] & 0.0 & [$-$2.3, $+$2.3] \\
 & dual & 144 & 129 & 89.6 [83.3, 95.1] & $-$0.8 & [$-$5.4, $+$3.9] \\
 & ZoomClick-noPZ & 144 & 126 & 87.5 [81.2, 93.1] & $+$0.8 & [$-$1.6, $+$4.0] \\
 & pooled (cat.) & 576 & 512 & 88.9 [82.6, 94.4] & $-$0.8 & [$-$2.6, $+$0.8] \\
\addlinespace
ScreenSpot-Pro, haiku$\to$sonnet, $\mathcal{C}$ & center & 48 & 38 & 79.2 [60.4, 91.7] & $+$2.6 & [0.0, $+$7.9] \\
 & grid & 48 & 35 & 72.9 [58.3, 83.3] & $-$5.7 & [$-$11.8, 0.0] \\
 & context & 48 & 36 & 75.0 [58.3, 89.6] & 0.0 & [0.0, 0.0] \\
 & dual & 48 & 37 & 77.1 [58.3, 91.7] & $+$8.1 & [0.0, $+$18.2] \\
 & ZoomClick-noPZ & 48 & 38 & 79.2 [62.5, 91.7] & 0.0 & [0.0, 0.0] \\
 & pooled (cat.) & 192 & 146 & 76.0 [59.4, 88.5] & $+$1.4 & [$-$1.5, $+$4.5] \\
\addlinespace
ScreenSpot-Pro, sonnet$\to$opus, $\mathcal{C}$ & center & 48 & 42 & 87.5 [66.7, 100.0] & $+$2.4 & [0.0, $+$7.1] \\
 & grid & 48 & 39 & 81.2 [58.3, 95.8] & $-$2.6 & [$-$7.3, 0.0] \\
 & context & 48 & 40 & 83.3 [58.3, 97.9] & 0.0 & [0.0, 0.0] \\
 & dual & 48 & 39 & 81.2 [56.2, 95.8] & $-$5.1 & [$-$17.9, $+$4.8] \\
 & ZoomClick-noPZ & 48 & 41 & 85.4 [60.4, 100.0] & 0.0 & [0.0, 0.0] \\
 & pooled (cat.) & 192 & 160 & 83.3 [60.9, 96.4] & $-$1.2 & [$-$4.3, $+$1.3] \\
\addlinespace
ScreenSpot-Pro, sonnet$\to$opus, $\mathcal{T}$ & center & 96 & 87 & 90.6 [84.4, 95.8] & $+$1.1 & [$-$2.3, $+$4.8] \\
 & grid & 96 & 79 & 82.3 [75.0, 88.5] & 0.0 & [$-$3.8, $+$3.7] \\
 & context & 96 & 86 & 89.6 [83.3, 94.8] & $-$2.3 & [$-$6.8, $+$2.2] \\
 & dual & 96 & 82 & 85.4 [78.1, 92.7] & $+$4.9 & [0.0, $+$10.0] \\
 & ZoomClick-noPZ & 96 & 82 & 85.4 [77.1, 91.7] & 0.0 & [0.0, 0.0] \\
 & pooled (cat.) & 384 & 334 & 87.0 [80.7, 92.2] & $+$0.9 & [$-$1.2, $+$2.8] \\
\bottomrule
\end{tabular}\end{adjustbox}

\caption{Agree/stable residual per catalogue skill and setting: tasks $n$, agree/stable tasks $N_{\mathrm{a/s}}$ and share, mean $F-R$ (points) with application-level bootstrap 95\% intervals.}
\label{tab:revagree}
\end{table}
\begin{table}[!htbp]
\centering
\scriptsize
\begin{adjustbox}{max width=\linewidth}\begin{tabular}{l r ccc cc}
\toprule
 & & \multicolumn{3}{c}{RCR minus competitor at RCR's request cost (interpolated)} & \multicolumn{2}{c}{RCR minus no-fresh policy} \\
\cmidrule(lr){3-5}\cmidrule(lr){6-7}
Setting & req. & uniform fresh & generic PPI & replay-PPI & replay-only & old decision \\
\midrule
UI-Vision, haiku$\to$sonnet & 257 & +0.1 [$-$0.1, +0.3] & +0.5 [+0.3, +0.7] & +0.1 [$-$0.1, +0.4] & $-$3.0 [$-$4.7, $-$1.5] & $-$3.0 [$-$4.7, $-$1.5] \\
UI-Vision, sonnet$\to$opus & 256 & $-$0.02 [$-$0.5, +0.5] & +0.1 [$-$0.4, +0.8] & +0.2 [+0.01, +0.4] & $-$0.4 [$-$1.1, +0.3] & +1.2 [+0.4, +2.1] \\
ScreenSpot-Pro, haiku$\to$sonnet & 129 & $-$0.2 [$-$0.4, $-$0.01] & $-$0.2 [$-$0.4, $-$0.01] & +0.9 [+0.1, +1.8] & $-$0.2 [$-$0.4, $-$0.01] & $-$0.2 [$-$0.4, $-$0.01] \\
ScreenSpot-Pro, sonnet$\to$opus & 129 & $-$0.8 [$-$1.3, $-$0.5] & $-$0.5 [$-$0.8, $-$0.3] & +1.0 [+0.5, +1.4] & $-$1.1 [$-$1.7, $-$0.7] & $-$1.1 [$-$1.7, $-$0.7] \\
\bottomrule
\end{tabular}\end{adjustbox}

\caption{Equal-cost comparison (interpolation preregistered, competitor set chosen post hoc): RCR's catalogue score minus each competitor's score linearly interpolated between its budget points to RCR's mean request cost per catalogue skill at the primary budget (``req.'', the anchor), and minus the two policies that use no fresh execution. Points, mean [app-level paired-bootstrap 95\% interval, conditional on the maintenance pool]. The two no-fresh columns are post hoc; exact cost matching is in Table~\ref{tab:reveqcost}.}
\label{tab:equalcost}
\end{table}
\paragraph{Equal request cost, matched exactly.} The preregistered comparison interpolates linearly between budget points (Table~\ref{tab:equalcost}). Table~\ref{tab:reveqcost} instead draws, per skill and history, an integer budget between the two that bracket RCR's request cost so that the expected cost equals it, over \cnt{rev.pol.n-hist} histories. RCR is then above uniform fresh in \cnt{rev.pol.eqc.uniform-fresh.n-ci-above-0} setting and below it in \cnt{rev.pol.eqc.uniform-fresh.n-ci-below-0}; most of the difference from Table~\ref{tab:equalcost} comes from replacing 20 histories by \cnt{rev.pol.n-hist}.

\begin{table}[!htbp]\centering\scriptsize
\begin{adjustbox}{max width=\linewidth}\begin{tabular}{l r rrr rrr rrr rrr}
\toprule
Setting & RCR req. & \multicolumn{3}{c}{uniform fresh} & \multicolumn{3}{c}{generic PPI} & \multicolumn{3}{c}{replay-PPI} & \multicolumn{3}{c}{best-single} \\
\cmidrule(lr){3-5}\cmidrule(lr){6-8}\cmidrule(lr){9-11}\cmidrule(lr){12-14}
 & & exact [95\% CI] & lin.$_N$ & lin.$_{20}$ & exact [95\% CI] & lin.$_N$ & lin.$_{20}$ & exact [95\% CI] & lin.$_N$ & lin.$_{20}$ & exact [95\% CI] & lin.$_N$ & lin.$_{20}$ \\
\midrule
UIV h$\to$s & 257 & +0.7 [+0.3, +1.2] & +0.9 & +0.1 & +0.01 [$-$0.1, +0.1] & +0.7 & +0.5 & +0.4 [+0.2, +0.7] & +0.4 & +0.1 & +1.6 [+0.4, +2.8] & $-$0.4 & $-$1.5 \\
UIV s$\to$o & 256 & +0.4 [$-$0.1, +1.0] & +0.2 & $-$0.02 & +0.2 [$-$0.02, +0.4] & $-$0.01 & +0.1 & +0.1 [+0.03, +0.2] & +0.1 & +0.2 & +0.5 [$-$0.5, +1.7] & +0.6 & +0.7 \\
SSP h$\to$s & 129 & $-$0.3 [$-$0.7, $-$0.05] & $-$0.3 & $-$0.2 & $-$0.4 [$-$0.8, $-$0.05] & $-$0.3 & $-$0.2 & +0.3 [+0.05, +0.7] & +0.3 & +0.9 & +0.4 [$-$1.6, +2.3] & +1.0 & +0.9 \\
SSP s$\to$o & 129 & $-$1.1 [$-$1.7, $-$0.6] & $-$1.1 & $-$0.8 & $-$1.1 [$-$1.7, $-$0.6] & $-$1.2 & $-$0.5 & +0.00 [$-$0.1, +0.1] & +0.00 & +1.0 & $-$1.8 [$-$3.0, $-$0.5] & $-$2.0 & $-$1.5 \\
\midrule
\quad realised req., UIV h$\to$s & & \multicolumn{3}{c}{257.1} & \multicolumn{3}{c}{257.1} & \multicolumn{3}{c}{257.1} & \multicolumn{3}{c}{257.1} \\
\quad realised req., UIV s$\to$o & & \multicolumn{3}{c}{256.0} & \multicolumn{3}{c}{256.0} & \multicolumn{3}{c}{256.0} & \multicolumn{3}{c}{256.0} \\
\quad realised req., SSP h$\to$s & & \multicolumn{3}{c}{128.9} & \multicolumn{3}{c}{128.9} & \multicolumn{3}{c}{128.9} & \multicolumn{3}{c}{128.9} \\
\quad realised req., SSP s$\to$o & & \multicolumn{3}{c}{128.8} & \multicolumn{3}{c}{128.8} & \multicolumn{3}{c}{128.8} & \multicolumn{3}{c}{128.8} \\
\bottomrule
\end{tabular}\end{adjustbox}

\caption{Exact equal-cost comparison over \cnt{rev.pol.n-hist} histories: RCR's mean request cost per catalogue skill and RCR minus each competitor (points; mean and application-level bootstrap 95\% interval).}
\label{tab:reveqcost}
\end{table}

\paragraph{Maintenance histories and the Holm family.} The tested quantity is the mean over 20 seeded histories, whose Monte Carlo standard error (\pct{rev.pol.holm.mc.min-se20}--\pct{rev.pol.holm.mc.max-se20} points) is as large as the differences. Re-running the frozen decision code with \cnt{rev.pol.n-hist} seeds reproduces the first 20 exactly and gives Table~\ref{tab:revholm}: \cnt{rev.pol.holm.n.n-rejected} of \cnt{holm.family-size} Holm rejections instead of \cnt{rev.pol.holm.f20.n-rejected}, with \cnt{rev.pol.holm.n-rejected-both} of the frozen ones surviving, \cnt{rev.pol.holm.n-lost-rejections} lost and \cnt{rev.pol.holm.n-new-rejections} new. No rejection depends on the bootstrap size or seed scheme, while exact sign-flip tests over applications (Monte Carlo with \cnt{rev.pol.signflip-mc} draws for \cnt{uiv.p1.n-test-apps} applications) and paired $t$-tests reject \cnt{rev.pol.perm.f20.signflip-gt.n-rejected} of \cnt{holm.family-size} with the frozen inputs and \cnt{rev.pol.perm.n.signflip-gt.n-rejected} with \cnt{rev.pol.n-hist} histories (Table~\ref{tab:revtests}). Non-inferiority to full recertification passes in \cnt{rev.pol.ni.n.n-pass} of \cnt{summary.all.ni-settings} settings with either input.

\begin{table}[!htbp]\centering\scriptsize
\begin{adjustbox}{max width=\linewidth}\begin{tabular}{l rrrrc rrrrc}
\toprule
 & \multicolumn{5}{c}{frozen{,} 20 histories} & \multicolumn{5}{c}{10{,}000 histories} \\
\cmidrule(lr){2-6}\cmidrule(lr){7-11}
Test & mean & LB & $p$ & Holm $p$ & rej. & mean & LB & $p$ & Holm $p$ & rej. \\
\midrule
\multicolumn{11}{l}{\emph{UI-Vision, haiku$\to$sonnet}} \\
\quad RCR $-$ replay-PPI & +0.1 & $-$0.1 & 0.165 & 0.165 & no & +0.4 & +0.2 & $<$0.001 & 0.002 & yes \\
\quad RCR $-$ generic PPI & +1.3 & +0.8 & $<$0.001 & 0.002 & yes & +0.8 & +0.4 & $<$0.001 & 0.002 & yes \\
\multicolumn{11}{l}{\emph{UI-Vision, sonnet$\to$opus}} \\
\quad RCR $-$ replay-PPI & +0.2 & +0.04 & 0.013 & 0.049 & yes & +0.1 & +0.04 & 0.003 & 0.021 & yes \\
\quad RCR $-$ generic PPI & +0.4 & +0.1 & 0.012 & 0.049 & yes & +0.1 & +0.00 & 0.046 & 0.137 & no \\
\multicolumn{11}{l}{\emph{ScreenSpot-Pro, haiku$\to$sonnet}} \\
\quad RCR $-$ replay-PPI & +0.9 & +0.2 & 0.014 & 0.049 & yes & +0.3 & +0.1 & 0.009 & 0.035 & yes \\
\quad RCR $-$ generic PPI & +0.6 & +0.2 & 0.004 & 0.021 & yes & +0.8 & +0.2 & 0.005 & 0.024 & yes \\
\multicolumn{11}{l}{\emph{ScreenSpot-Pro, sonnet$\to$opus}} \\
\quad RCR $-$ replay-PPI & +1.0 & +0.6 & $<$0.001 & 0.002 & yes & +0.01 & $-$0.04 & 0.438 & 0.876 & no \\
\quad RCR $-$ generic PPI & +1.0 & +0.6 & $<$0.001 & 0.002 & yes & $-$0.01 & $-$0.1 & 0.577 & 0.876 & no \\
\midrule
Rejected (of 8) & \multicolumn{5}{c}{7} & \multicolumn{5}{c}{5} \\
\bottomrule
\end{tabular}\end{adjustbox}

\caption{The frozen Holm family with the 20 frozen histories and with \cnt{rev.pol.n-hist} histories: mean difference (points), one-sided 95\% lower bound, raw and Holm-adjusted $p$, rejection.}
\label{tab:revholm}
\end{table}

\begin{table}[!htbp]\centering\scriptsize
\begin{adjustbox}{max width=\linewidth}\begin{tabular}{l rrr rrr rr}
\toprule
 & \multicolumn{3}{c}{Holm $p${,} $B=100{,}000${,} frozen inputs} & \multicolumn{3}{c}{Holm $p${,} $B=100{,}000${,} $N$ inputs} & \multicolumn{2}{c}{Holm $p${,} frozen inputs} \\
\cmidrule(lr){2-4}\cmidrule(lr){5-7}\cmidrule(lr){8-9}
Test & content & fixed A & fixed B & content & fixed A & fixed B & sign-flip & $t$ \\
\midrule
\multicolumn{9}{l}{\emph{UI-Vision, haiku$\to$sonnet}} \\
\quad RCR $-$ replay-PPI & 0.164 & 0.165 & 0.164 & $<$0.001$^{*}$ & $<$0.001$^{*}$ & $<$0.001$^{*}$ & 0.232 & 0.182 \\
\quad RCR $-$ generic PPI & $<$0.001$^{*}$ & $<$0.001$^{*}$ & $<$0.001$^{*}$ & $<$0.001$^{*}$ & $<$0.001$^{*}$ & $<$0.001$^{*}$ & 0.002$^{*}$ & 0.004$^{*}$ \\
\multicolumn{9}{l}{\emph{UI-Vision, sonnet$\to$opus}} \\
\quad RCR $-$ replay-PPI & 0.047$^{*}$ & 0.047$^{*}$ & 0.046$^{*}$ & 0.019$^{*}$ & 0.020$^{*}$ & 0.017$^{*}$ & 0.145 & 0.134 \\
\quad RCR $-$ generic PPI & 0.047$^{*}$ & 0.047$^{*}$ & 0.046$^{*}$ & 0.134 & 0.134 & 0.130 & 0.145 & 0.134 \\
\multicolumn{9}{l}{\emph{ScreenSpot-Pro, haiku$\to$sonnet}} \\
\quad RCR $-$ replay-PPI & 0.047$^{*}$ & 0.047$^{*}$ & 0.046$^{*}$ & 0.036$^{*}$ & 0.036$^{*}$ & 0.037$^{*}$ & 0.145 & 0.134 \\
\quad RCR $-$ generic PPI & 0.027$^{*}$ & 0.027$^{*}$ & 0.028$^{*}$ & 0.027$^{*}$ & 0.027$^{*}$ & 0.028$^{*}$ & 0.145 & 0.134 \\
\multicolumn{9}{l}{\emph{ScreenSpot-Pro, sonnet$\to$opus}} \\
\quad RCR $-$ replay-PPI & $<$0.001$^{*}$ & $<$0.001$^{*}$ & $<$0.001$^{*}$ & 0.889 & 0.891 & 0.893 & 0.007$^{*}$ & 0.006$^{*}$ \\
\quad RCR $-$ generic PPI & $<$0.001$^{*}$ & $<$0.001$^{*}$ & $<$0.001$^{*}$ & 0.889 & 0.891 & 0.893 & 0.007$^{*}$ & 0.006$^{*}$ \\
\midrule
Rejected (of 8) & 7 & 7 & 7 & 5 & 5 & 5 & 3 & 3 \\
\bottomrule
\end{tabular}\end{adjustbox}

\caption{Holm-adjusted $p$ for the same family under $B=\cnt{rev.pol.boot-B-hi}$ bootstrap resamples with three seed schemes, and under application-level sign-flip and paired $t$-tests (frozen inputs); with \cnt{rev.pol.n-hist}-history inputs these reject \cnt{rev.pol.perm.n.signflip-gt.n-rejected} and \cnt{rev.pol.perm.n.ttest-gt.n-rejected} of \cnt{holm.family-size}.}
\label{tab:revtests}
\end{table}

\paragraph{Minimum detectable differences.} With $G$ applications and $s_d$ the SD of the application-level paired differences, MDD $=(t_{0.95,G-1}+t_{0.80,G-1})\,s_d/\sqrt{G}$ (one-sided $\alpha=0.05$, power 0.8; Table~\ref{tab:revmdd}). For the superiority comparisons the MDD is \pct{rev.pol.mdd.f20.sup.min}--\pct{rev.pol.mdd.f20.sup.max} points and \cnt{rev.pol.mdd.f20.sup.n-obs-above-mdd} of \cnt{holm.family-size} observed differences reach it (a statement about power, not about the tests); for a skill's sealed effect it is \pct{rev.pol.mdd-tau.uiv.min}--\pct{rev.pol.mdd-tau.uiv.max} points on UI-Vision and \pct{rev.pol.mdd-tau.sspro.min}--\pct{rev.pol.mdd-tau.sspro.max} on ScreenSpot-Pro. The MDD is conditional on the pool and ignores history noise.

\begin{table}[!htbp]\centering\scriptsize
\begin{adjustbox}{max width=\linewidth}\begin{tabular}{l l r r r r r}
\toprule
Setting & Comparison / skill & $G$ & $s_d$ & observed & MDD$_{0.05}$ & MDD$_{0.05/8}$ \\
\midrule
\multicolumn{7}{l}{\emph{Policy comparisons at the primary budget (frozen 20-history inputs; points)}} \\
UIV h$\to$s & RCR $-$ replay-PPI & 24 & 0.7 & +0.1 & 0.4 & 0.5 \\
UIV h$\to$s & RCR $-$ generic PPI & 24 & 1.7 & +1.3 & 0.9 & 1.3 \\
UIV h$\to$s & RCR $-$ full recert.\ (NI) & 24 & 2.0 & +1.0 & 1.1 & 1.5 \\
UIV s$\to$o & RCR $-$ replay-PPI & 24 & 0.4 & +0.2 & 0.2 & 0.3 \\
UIV s$\to$o & RCR $-$ generic PPI & 24 & 0.9 & +0.4 & 0.5 & 0.6 \\
UIV s$\to$o & RCR $-$ full recert.\ (NI) & 24 & 1.9 & $-$0.2 & 1.0 & 1.4 \\
SSP h$\to$s & RCR $-$ replay-PPI & 12 & 1.6 & +0.9 & 1.2 & 1.8 \\
SSP h$\to$s & RCR $-$ generic PPI & 12 & 1.0 & +0.6 & 0.8 & 1.2 \\
SSP h$\to$s & RCR $-$ full recert.\ (NI) & 12 & 0.3 & $-$0.2 & 0.3 & 0.4 \\
SSP s$\to$o & RCR $-$ replay-PPI & 12 & 0.8 & +1.0 & 0.6 & 0.9 \\
SSP s$\to$o & RCR $-$ generic PPI & 12 & 0.8 & +1.0 & 0.6 & 0.9 \\
SSP s$\to$o & RCR $-$ full recert.\ (NI) & 12 & 1.0 & $-$1.1 & 0.8 & 1.1 \\
\midrule
\multicolumn{7}{l}{\emph{Per-skill sealed effect $\tau$ on $\mathcal{T}$ (catalogue skills; points)}} \\
UIV h$\to$s & center & 24 & 15.3 & $-$5.6 & 8.0 & 11.1 \\
UIV h$\to$s & grid & 24 & 15.5 & $-$7.6 & 8.1 & 11.3 \\
UIV h$\to$s & context & 24 & 13.6 & $-$5.6 & 7.1 & 9.9 \\
UIV h$\to$s & dual & 24 & 13.7 & $-$10.4 & 7.2 & 10.0 \\
UIV s$\to$o & center & 24 & 10.6 & $-$2.8 & 5.6 & 7.7 \\
UIV s$\to$o & grid & 24 & 9.2 & $-$0.7 & 4.8 & 6.7 \\
UIV s$\to$o & context & 24 & 7.0 & 0.0 & 3.6 & 5.1 \\
UIV s$\to$o & dual & 24 & 15.4 & $-$6.2 & 8.1 & 11.2 \\
SSP h$\to$s & center & 12 & 19.1 & +10.4 & 14.7 & 21.3 \\
SSP h$\to$s & grid & 12 & 21.6 & +11.5 & 16.7 & 24.1 \\
SSP h$\to$s & context & 12 & 22.3 & +11.5 & 17.2 & 24.8 \\
SSP h$\to$s & dual & 12 & 27.1 & +14.6 & 20.9 & 30.2 \\
SSP s$\to$o & center & 12 & 14.2 & +15.6 & 11.0 & 15.8 \\
SSP s$\to$o & grid & 12 & 15.8 & +14.6 & 12.2 & 17.6 \\
SSP s$\to$o & context & 12 & 14.1 & +12.5 & 10.9 & 15.7 \\
SSP s$\to$o & dual & 12 & 13.2 & +15.6 & 10.2 & 14.7 \\
\bottomrule
\end{tabular}\end{adjustbox}

\caption{Minimum detectable differences at the realised application counts $G$ for the preregistered comparisons and the per-skill sealed effects; observed differences for reference (points).}
\label{tab:revmdd}
\end{table}

\paragraph{Decision propensity.} Table~\ref{tab:revkeep} gives each budgeted policy's KEEP rate and score. The sealed oracle keeps \pct{rev.pol.keep.uiv.p1.oracle-on-T.keepN}\%, \pct{rev.pol.keep.uiv.p2.oracle-on-T.keepN}\%, \pct{rev.pol.keep.sspro.p1.oracle-on-T.keepN}\% and \pct{rev.pol.keep.sspro.p2.oracle-on-T.keepN}\% of the catalogue in the four settings. Where it is constant across skills, score differences arise only through which skills are kept, all in one direction, and the correlation over the \cnt{rev.pol.keep.uiv.p1.corr.budgeted-points} budgeted points is $\val{rev.pol.keep.uiv.p1.corr.pearson}$ (UI-Vision haiku$\to$sonnet), $\val{rev.pol.keep.sspro.p1.corr.pearson}$ and $\val{rev.pol.keep.sspro.p2.corr.pearson}$ (ScreenSpot-Pro); where it is mixed (UI-Vision sonnet$\to$opus) it is $\val{rev.pol.keep.uiv.p2.corr.pearson}$. The correlation therefore cannot separate a directional bias from better decisions. At the primary budget RCR keeps the fewest skills where the oracle suspends all (\pct{rev.pol.keep.uiv.p1.rcr-stratified@0.25.keepN}\%, UI-Vision haiku$\to$sonnet) and the most on ScreenSpot-Pro haiku$\to$sonnet (\pct{rev.pol.keep.sspro.p1.rcr-stratified@0.25.keepN}\%), but on ScreenSpot-Pro sonnet$\to$opus it keeps \pct{rev.pol.keep.sspro.p2.rcr-stratified@0.25.keepN}\% against \pct{rev.pol.keep.sspro.p2.uniform-fresh@0.25.keepN}\% for uniform fresh; the pattern also differs at other budgets. Better decisions at equal fresh budget are therefore possible but not established, and at equal request cost RCR is ahead of uniform fresh only on UI-Vision haiku$\to$sonnet (Table~\ref{tab:reveqcost}).

\begin{table}[!htbp]\centering\scriptsize
\begin{adjustbox}{max width=\linewidth}\begin{tabular}{l rrr rrr rrr rrr}
\toprule
Policy@budget & \multicolumn{3}{c}{UIV h$\to$s} & \multicolumn{3}{c}{UIV s$\to$o} & \multicolumn{3}{c}{SSP h$\to$s} & \multicolumn{3}{c}{SSP s$\to$o} \\
\cmidrule(lr){2-4}\cmidrule(lr){5-7}\cmidrule(lr){8-10}\cmidrule(lr){11-13}
 & K$_{20}$ & K$_N$ & score$_N$ & K$_{20}$ & K$_N$ & score$_N$ & K$_{20}$ & K$_N$ & score$_N$ & K$_{20}$ & K$_N$ & score$_N$ \\
\midrule
uniform fresh@10\% & 58.8 & 66.8 & 78.5 & 83.8 & 77.8 & 82.2 & 90.0 & 91.6 & 84.9 & 92.5 & 91.6 & 89.4 \\
replay-PPI@10\% & 40.0 & 43.0 & 80.1 & 61.3 & 60.1 & 82.9 & 90.0 & 86.5 & 84.3 & 88.8 & 89.1 & 89.0 \\
generic PPI@10\% & 55.0 & 51.4 & 79.8 & 52.5 & 46.1 & 83.1 & 91.2 & 87.9 & 84.5 & 92.5 & 91.5 & 89.3 \\
RCR@10\% & 33.8 & 38.3 & 80.5 & 51.2 & 58.2 & 82.6 & 0.0 & 0.0 & 74.0 & 66.2 & 66.0 & 85.6 \\
\addlinespace
uniform fresh@25\% & 50.0 & 57.0 & 79.2 & 73.8 & 71.8 & 82.4 & 92.5 & 94.9 & 85.3 & 90.0 & 92.5 & 89.5 \\
replay-PPI@25\% & 41.2 & 43.5 & 80.0 & 50.0 & 58.0 & 82.7 & 92.5 & 94.9 & 85.2 & 86.2 & 89.7 & 89.1 \\
generic PPI@25\% & 57.5 & 49.6 & 79.7 & 55.0 & 54.4 & 82.7 & 93.8 & 90.1 & 84.7 & 86.2 & 89.8 & 89.1 \\
RCR@25\% & 38.8 & 37.9 & 80.5 & 48.8 & 58.8 & 82.8 & 98.8 & 97.3 & 85.6 & 92.5 & 89.7 & 89.1 \\
\addlinespace
uniform fresh@50\% & 38.8 & 50.4 & 79.6 & 65.0 & 72.6 & 82.5 & 100.0 & 99.2 & 85.8 & 97.5 & 96.3 & 90.1 \\
replay-PPI@50\% & 32.5 & 41.4 & 80.3 & 67.5 & 68.8 & 82.6 & 100.0 & 99.8 & 85.9 & 93.8 & 97.1 & 90.2 \\
generic PPI@50\% & 43.8 & 45.5 & 79.9 & 66.2 & 63.5 & 82.7 & 100.0 & 99.0 & 85.8 & 95.0 & 97.1 & 90.2 \\
RCR@50\% & 46.2 & 41.3 & 80.4 & 42.5 & 60.3 & 82.9 & 100.0 & 99.7 & 85.9 & 97.5 & 96.1 & 90.0 \\
\addlinespace
uniform fresh@100\% & 50.0 & 50.0 & 79.3 & 75.0 & 75.0 & 83.2 & 100.0 & 100.0 & 85.9 & 100.0 & 100.0 & 90.6 \\
replay-PPI@100\% & 50.0 & 50.0 & 79.3 & 75.0 & 75.0 & 83.2 & 100.0 & 100.0 & 85.9 & 100.0 & 100.0 & 90.6 \\
generic PPI@100\% & 50.0 & 50.0 & 79.3 & 75.0 & 75.0 & 83.2 & 100.0 & 100.0 & 85.9 & 100.0 & 100.0 & 90.6 \\
RCR@100\% & 50.0 & 50.0 & 79.3 & 75.0 & 75.0 & 83.2 & 100.0 & 100.0 & 85.9 & 100.0 & 100.0 & 90.6 \\
\addlinespace
keep-all & 100.0 & 100.0 & 76.0 & 100.0 & 100.0 & 81.6 & 100.0 & 100.0 & 85.9 & 100.0 & 100.0 & 90.6 \\
suspend-all & 0.0 & 0.0 & 83.3 & 0.0 & 0.0 & 84.0 & 0.0 & 0.0 & 74.0 & 0.0 & 0.0 & 76.0 \\
old decision & 0.0 & 0.0 & 83.3 & 50.0 & 50.0 & 81.8 & 100.0 & 100.0 & 85.9 & 100.0 & 100.0 & 90.6 \\
replay-only & 0.0 & 0.0 & 83.3 & 50.0 & 50.0 & 83.3 & 100.0 & 100.0 & 85.9 & 100.0 & 100.0 & 90.6 \\
full recert. & 50.0 & 50.0 & 79.3 & 75.0 & 75.0 & 83.2 & 100.0 & 100.0 & 85.9 & 100.0 & 100.0 & 90.6 \\
oracle on $\mathcal{T}$ & 0.0 & 0.0 & 83.3 & 25.0 & 25.0 & 84.0 & 100.0 & 100.0 & 85.9 & 100.0 & 100.0 & 90.6 \\
\midrule
Pearson $r$ (16 budgeted points) & \multicolumn{3}{c}{$-$0.95} & \multicolumn{3}{c}{$-$0.08} & \multicolumn{3}{c}{+1.00} & \multicolumn{3}{c}{+1.00} \\
Spearman $\rho$ (16 budgeted points) & \multicolumn{3}{c}{$-$0.90} & \multicolumn{3}{c}{+0.11} & \multicolumn{3}{c}{+1.00} & \multicolumn{3}{c}{+1.00} \\
Spearman $\rho$ (incl.\ no-budget, 21) & \multicolumn{3}{c}{$-$0.94} & \multicolumn{3}{c}{$-$0.16} & \multicolumn{3}{c}{+1.00} & \multicolumn{3}{c}{+1.00} \\
\bottomrule
\end{tabular}\end{adjustbox}

\caption{KEEP rate of each budgeted policy (\%, over the 20 frozen histories, K$_{20}$, and \cnt{rev.pol.n-hist} histories, K$_N$) and its catalogue score over $N$ histories.}
\label{tab:revkeep}
\end{table}

\paragraph{Margin and cost-tied decision rules.} Table~\ref{tab:revrules} re-decides every policy with KEEP iff $\hat\tau\ge m$ ($m=0.5,1,2$ points) or KEEP iff $\hat\tau\ge\lambda(r_s-1)$ points, with $r_s$ the skill's requests per task on $\mathcal{C}$ ($\lambda=0.5,1$), and reports the deployment requests per task on $\mathcal{T}$ that each decision commits to. Margins help on UI-Vision (up to \ppt{rev.pol.rules.gain.uiv.m2.max} points) and hurt on ScreenSpot-Pro (down to \ppt{rev.pol.rules.gain.sspro.m2.min}); no rule improves on the frozen one in every setting. Table~\ref{tab:revrulekeep} gives the share of catalogue skills each rule keeps.

\begin{table}[!htbp]\centering\tiny
\begin{adjustbox}{max width=\linewidth}\begin{tabular}{ll rr rr rr rr}
\toprule
Policy & Rule & \multicolumn{2}{c}{UIV h$\to$s} & \multicolumn{2}{c}{UIV s$\to$o} & \multicolumn{2}{c}{SSP h$\to$s} & \multicolumn{2}{c}{SSP s$\to$o} \\
\cmidrule(lr){3-4}\cmidrule(lr){5-6}\cmidrule(lr){7-8}\cmidrule(lr){9-10}
 & & score & req. & score & req. & score & req. & score & req. \\
\midrule
uniform fresh & frozen ($m=0$) & 79.2 & 1.91 & 82.4 & 2.16 & 85.3 & 2.63 & 89.5 & 2.52 \\
 & $m=0.5$ & 81.1 & 1.49 & 83.2 & 1.62 & 83.5 & 2.37 & 86.3 & 2.16 \\
 & $m=1$ & 81.1 & 1.49 & 83.2 & 1.62 & 83.5 & 2.37 & 86.3 & 2.16 \\
 & $m=2$ & 81.1 & 1.49 & 83.2 & 1.62 & 83.5 & 2.37 & 86.3 & 2.16 \\
 & $\lambda=0.5$ & 81.1 & 1.49 & 83.2 & 1.62 & 83.5 & 2.37 & 86.3 & 2.16 \\
 & $\lambda=1$ & 81.1 & 1.49 & 83.2 & 1.62 & 83.5 & 2.37 & 86.3 & 2.16 \\
\addlinespace
replay-PPI & frozen ($m=0$) & 80.0 & 1.73 & 82.7 & 1.91 & 85.2 & 2.62 & 89.1 & 2.45 \\
 & $m=0.5$ & 80.0 & 1.73 & 82.7 & 1.91 & 85.2 & 2.62 & 89.1 & 2.45 \\
 & $m=1$ & 80.0 & 1.73 & 82.7 & 1.91 & 85.2 & 2.62 & 89.1 & 2.45 \\
 & $m=2$ & 81.1 & 1.47 & 83.0 & 1.82 & 85.2 & 2.62 & 89.1 & 2.45 \\
 & $\lambda=0.5$ & 80.4 & 1.61 & 82.7 & 1.91 & 85.2 & 2.62 & 89.1 & 2.45 \\
 & $\lambda=1$ & 81.1 & 1.47 & 82.8 & 1.75 & 85.2 & 2.62 & 89.1 & 2.45 \\
\addlinespace
generic PPI & frozen ($m=0$) & 79.7 & 1.80 & 82.7 & 1.89 & 84.7 & 2.58 & 89.1 & 2.45 \\
 & $m=0.5$ & 80.2 & 1.69 & 82.7 & 1.89 & 84.7 & 2.58 & 89.1 & 2.45 \\
 & $m=1$ & 80.2 & 1.69 & 82.7 & 1.89 & 84.7 & 2.58 & 89.1 & 2.45 \\
 & $m=2$ & 80.8 & 1.59 & 83.2 & 1.74 & 84.7 & 2.58 & 89.1 & 2.45 \\
 & $\lambda=0.5$ & 80.2 & 1.69 & 82.7 & 1.89 & 84.7 & 2.58 & 89.1 & 2.45 \\
 & $\lambda=1$ & 80.5 & 1.59 & 83.2 & 1.59 & 84.7 & 2.58 & 89.1 & 2.45 \\
\addlinespace
RCR & frozen ($m=0$) & 80.5 & 1.63 & 82.8 & 1.91 & 85.6 & 2.66 & 89.1 & 2.44 \\
 & $m=0.5$ & 80.7 & 1.59 & 83.0 & 1.77 & 85.6 & 2.66 & 88.5 & 2.40 \\
 & $m=1$ & 80.7 & 1.57 & 83.0 & 1.71 & 85.1 & 2.60 & 88.5 & 2.40 \\
 & $m=2$ & 81.4 & 1.42 & 83.1 & 1.61 & 85.1 & 2.60 & 88.1 & 2.34 \\
 & $\lambda=0.5$ & 80.8 & 1.54 & 83.0 & 1.77 & 85.1 & 2.60 & 88.5 & 2.40 \\
 & $\lambda=1$ & 81.2 & 1.45 & 83.0 & 1.66 & 84.8 & 2.52 & 88.1 & 2.34 \\
\addlinespace
replay-only & frozen ($m=0$) & 83.3 & 1.00 & 83.3 & 1.50 & 85.9 & 2.72 & 90.6 & 2.67 \\
 & $m=0.5$ & 83.3 & 1.00 & 83.3 & 1.50 & 85.9 & 2.72 & 90.6 & 2.67 \\
 & $m=1$ & 83.3 & 1.00 & 83.3 & 1.50 & 85.9 & 2.72 & 90.6 & 2.67 \\
 & $m=2$ & 83.3 & 1.00 & 84.0 & 1.25 & 85.9 & 2.72 & 90.6 & 2.67 \\
 & $\lambda=0.5$ & 83.3 & 1.00 & 83.3 & 1.50 & 85.9 & 2.72 & 90.6 & 2.67 \\
 & $\lambda=1$ & 83.3 & 1.00 & 83.3 & 1.50 & 85.9 & 2.72 & 90.6 & 2.67 \\
\addlinespace
full recert. & frozen ($m=0$) & 79.3 & 1.76 & 83.2 & 2.15 & 85.9 & 2.72 & 90.6 & 2.67 \\
 & $m=0.5$ & 81.9 & 1.25 & 83.2 & 2.15 & 85.9 & 2.72 & 90.6 & 2.67 \\
 & $m=1$ & 81.9 & 1.25 & 83.2 & 2.15 & 85.9 & 2.72 & 90.6 & 2.67 \\
 & $m=2$ & 83.3 & 1.00 & 83.3 & 1.25 & 85.9 & 2.72 & 90.6 & 2.67 \\
 & $\lambda=0.5$ & 81.9 & 1.25 & 83.3 & 1.50 & 85.9 & 2.72 & 90.6 & 2.67 \\
 & $\lambda=1$ & 81.9 & 1.25 & 83.3 & 1.50 & 85.9 & 2.72 & 90.6 & 2.67 \\
\addlinespace
old decision & frozen ($m=0$) & 83.3 & 1.00 & 81.8 & 1.75 & 85.9 & 2.72 & 90.6 & 2.67 \\
 & $m=0.5$ & 83.3 & 1.00 & 83.3 & 1.25 & 85.9 & 2.72 & 90.6 & 2.67 \\
 & $m=1$ & 83.3 & 1.00 & 83.3 & 1.25 & 85.9 & 2.72 & 90.6 & 2.67 \\
 & $m=2$ & 83.3 & 1.00 & 84.0 & 1.00 & 85.9 & 2.72 & 90.6 & 2.67 \\
 & $\lambda=0.5$ & 83.3 & 1.00 & 83.3 & 1.25 & 85.9 & 2.72 & 90.6 & 2.67 \\
 & $\lambda=1$ & 83.3 & 1.00 & 83.3 & 1.25 & 85.9 & 2.72 & 90.6 & 2.67 \\
\addlinespace
suspend-all & -- & 83.3 & 1.00 & 84.0 & 1.00 & 74.0 & 1.00 & 76.0 & 1.00 \\
keep-all & -- & 76.0 & 2.67 & 81.6 & 2.65 & 85.9 & 2.72 & 90.6 & 2.67 \\
oracle on $\mathcal{T}$ & -- & 83.3 & 1.00 & 84.0 & 1.25 & 85.9 & 2.72 & 90.6 & 2.67 \\
\bottomrule
\end{tabular}\end{adjustbox}

\caption{Exploratory decision rules: catalogue score (\%) and deployment requests per task on $\mathcal{T}$ under the frozen rule ($m=0$), margin rules and cost-tied rules, per policy and setting ($N$ histories for budgeted policies).}
\label{tab:revrules}
\end{table}
\begin{table}[!htbp]\centering\scriptsize
\begin{adjustbox}{max width=\linewidth}\begin{tabular}{ll rrrr}
\toprule
Policy & Rule & UIV h$\to$s & UIV s$\to$o & SSP h$\to$s & SSP s$\to$o \\
\midrule
old decision & $m=0$ (frozen) & 0 & 50 & 100 & 100 \\
 & $m=0.5$ & 0 & 25 & 100 & 100 \\
 & $m=1$ & 0 & 25 & 100 & 100 \\
 & $m=2$ & 0 & 0 & 100 & 100 \\
 & $\lambda=0.5$ & 0 & 25 & 100 & 100 \\
 & $\lambda=1$ & 0 & 25 & 100 & 100 \\
\addlinespace
replay-only & $m=0$ (frozen) & 0 & 50 & 100 & 100 \\
 & $m=0.5$ & 0 & 50 & 100 & 100 \\
 & $m=1$ & 0 & 50 & 100 & 100 \\
 & $m=2$ & 0 & 25 & 100 & 100 \\
 & $\lambda=0.5$ & 0 & 50 & 100 & 100 \\
 & $\lambda=1$ & 0 & 50 & 100 & 100 \\
\addlinespace
full recert. & $m=0$ (frozen) & 50 & 75 & 100 & 100 \\
 & $m=0.5$ & 25 & 75 & 100 & 100 \\
 & $m=1$ & 25 & 75 & 100 & 100 \\
 & $m=2$ & 0 & 25 & 100 & 100 \\
 & $\lambda=0.5$ & 25 & 50 & 100 & 100 \\
 & $\lambda=1$ & 25 & 50 & 100 & 100 \\
\addlinespace
uniform fresh & $m=0$ (frozen) & 57 & 72 & 95 & 93 \\
 & $m=0.5$ & 31 & 40 & 80 & 71 \\
 & $m=1$ & 31 & 40 & 80 & 71 \\
 & $m=2$ & 31 & 40 & 80 & 71 \\
 & $\lambda=0.5$ & 31 & 40 & 80 & 71 \\
 & $\lambda=1$ & 31 & 40 & 80 & 71 \\
\addlinespace
generic PPI & $m=0$ (frozen) & 50 & 54 & 90 & 90 \\
 & $m=0.5$ & 44 & 54 & 90 & 90 \\
 & $m=1$ & 44 & 54 & 90 & 90 \\
 & $m=2$ & 34 & 47 & 90 & 90 \\
 & $\lambda=0.5$ & 44 & 54 & 90 & 90 \\
 & $\lambda=1$ & 40 & 41 & 90 & 90 \\
\addlinespace
replay-PPI & $m=0$ (frozen) & 44 & 58 & 95 & 90 \\
 & $m=0.5$ & 44 & 58 & 95 & 90 \\
 & $m=1$ & 44 & 58 & 95 & 90 \\
 & $m=2$ & 32 & 49 & 95 & 90 \\
 & $\lambda=0.5$ & 39 & 58 & 95 & 90 \\
 & $\lambda=1$ & 32 & 52 & 95 & 90 \\
\addlinespace
RCR & $m=0$ (frozen) & 38 & 59 & 97 & 90 \\
 & $m=0.5$ & 36 & 53 & 97 & 86 \\
 & $m=1$ & 34 & 47 & 94 & 86 \\
 & $m=2$ & 24 & 40 & 94 & 83 \\
 & $\lambda=0.5$ & 34 & 53 & 94 & 86 \\
 & $\lambda=1$ & 29 & 45 & 92 & 83 \\
\bottomrule
\end{tabular}\end{adjustbox}

\caption{Share of catalogue skills kept (\%) under the frozen rule, margin rules and cost-tied rules (budgeted policies at the primary budget, $N$ histories).}
\label{tab:revrulekeep}
\end{table}

\paragraph{Triggered allocation.} The preregistered policies spend a fixed fraction of $|\mathcal{C}|$ per skill. Table~\ref{tab:revtrig} evaluates a triggered alternative that replays every task, marks a skill as in doubt if its replay estimate lies within two standard errors of zero (\cnt{rev.pol.trig.se.n-doubt-total} of \cnt{rev.pol.trig.n-skill-settings} skill-settings), or alternatively if replay and the old decision disagree (\cnt{rev.pol.trig.disagree.n-doubt-total}), and spends the same total fresh budget only on those skills with the replay-PPI correction. Neither variant scores above replay-only in any setting.

\begin{table}[!htbp]\centering\scriptsize
\begin{adjustbox}{max width=\linewidth}\begin{tabular}{ll rrrr rrr}
\toprule
Setting & in doubt & score (SD) & KEEP & req. & $\Delta$req. & $-$RCR & $-$replay-PPI & $-$replay-only \\
\midrule
\multicolumn{9}{l}{\emph{in doubt if $|\hat\tau_{\mathrm{rep}}|<2\,\mathrm{SE}$}} \\
\quad UIV h$\to$s & center, context, dual & 80.9 (1.7) & 34.7 & 248 & $-$9 & +0.4 [+0.1, +0.7] & +0.8 [+0.3, +1.3] & $-$2.5 [$-$3.9, $-$1.2] \\
\quad UIV s$\to$o & center, grid, context, dual & 82.7 (0.9) & 58.0 & 256 & $-$1 & $-$0.1 [$-$0.2, $-$0.02] & 0.0 [0.0, 0.0] & $-$0.6 [$-$1.4, +0.2] \\
\quad SSP h$\to$s & center, grid, context, dual & 85.2 (1.5) & 94.9 & 129 & 0 & $-$0.3 [$-$0.7, $-$0.03] & 0.0 [0.0, 0.0] & $-$0.7 [$-$1.4, $-$0.1] \\
\quad SSP s$\to$o & center, grid, dual & 89.4 (1.8) & 92.4 & 132 & +3 & +0.4 [+0.1, +0.7] & +0.4 [+0.2, +0.6] & $-$1.2 [$-$1.7, $-$0.6] \\
\multicolumn{9}{l}{\emph{in doubt if replay $\ne$ old decision}} \\
\quad UIV h$\to$s & none & 83.3 (0.0) & 0.0 & 192 & $-$65 & +2.9 [+1.4, +4.4] & +3.3 [+1.6, +5.1] & 0.0 [0.0, 0.0] \\
\quad UIV s$\to$o & context, dual & 82.6 (0.8) & 56.7 & 252 & $-$4 & $-$0.3 [$-$0.8, +0.2] & $-$0.2 [$-$0.6, +0.3] & $-$0.8 [$-$1.5, $-$0.1] \\
\quad SSP h$\to$s & none & 85.9 (0.0) & 100.0 & 96 & $-$33 & +0.4 [+0.03, +0.7] & +0.7 [+0.1, +1.4] & 0.0 [0.0, 0.0] \\
\quad SSP s$\to$o & none & 90.6 (0.0) & 100.0 & 96 & $-$33 & +1.6 [+0.9, +2.3] & +1.6 [+0.9, +2.3] & 0.0 [0.0, 0.0] \\
\bottomrule
\end{tabular}\end{adjustbox}

\caption{Exploratory triggered allocation: skills in doubt, catalogue score (SD), KEEP rate, request cost per skill and its change, and differences to RCR, replay-PPI and replay-only (points; $N$ histories).}
\label{tab:revtrig}
\end{table}

\paragraph{Zero-read answers.} Treating the \cnt{rev.pol.zr.n-deferred} zero-read answers that remain in the frozen data as missing changes \cnt{rev.pol.zr.missing.n-changed} of \cnt{rev.pol.zr.missing.n-total} budgeted decisions, mostly by re-sampling (the frozen samplers draw from the reduced pool), and any catalogue score by at most \pct{rev.pol.zr.missing.max-abs-score-delta} points; treating them as wrong changes \cnt{rev.pol.zr.incorrect.n-changed} decisions and scores by at most \pct{rev.pol.zr.incorrect.max-abs-score-delta} points.

\paragraph{Material flips and cross-model changes.} Table~\ref{tab:revdtau} gives, per skill, the cross-model change $\Delta\hat\tau=\hat\tau_1-\hat\tau_0$ with an application-level interval and equivalence tests at $\pm1$ and $\pm5$ points, on $\mathcal{T}$ for sonnet$\to$opus and on $\mathcal{C}$ or across pools for haiku$\to$sonnet (haiku's workflow was not run on $\mathcal{T}$). Under the definition of Sec.~4.2 there are \cnt{rev.desc.summary.d1.same-T.material-flips}, \cnt{rev.desc.summary.d1.same-C.material-flips} and \cnt{rev.desc.summary.d1.cross.material-flips} material flips on the three bases; equivalence within $\pm5$ points could not be established for any skill (\cnt{rev.desc.summary.d1.same-T.tost5-equivalent}, \cnt{rev.desc.summary.d1.same-C.tost5-equivalent}, \cnt{rev.desc.summary.d1.cross.tost5-equivalent}). Table~\ref{tab:revflips} gives per-task positive and negative flips between the two models' workflows and direct clicks, and between repeated direct runs of one model.

\begin{table}[!htbp]\centering\scriptsize
\begin{adjustbox}{max width=\linewidth}\begin{tabular}{ll l rrr l cc}
\toprule
Domain, change & Skill & basis & $\hat\tau_0$ & $\hat\tau_1$ & $\Delta\hat\tau$ & 95\% CI & TOST $\pm1$ & TOST $\pm5$ \\
\midrule
UI-Vision, haiku$\to$sonnet & center & $\mathcal{C}$ & $-$7.3 & $+$1.0 & $+$8.3 & [$-$4.2, $+$19.8] & no & no \\
 & grid & $\mathcal{C}$ & $-$14.6 & $-$3.1 & $+$11.5 & [$+$3.1, $+$19.8] & no & no \\
 & context & $\mathcal{C}$ & $-$7.3 & $-$2.1 & $+$5.2 & [$-$5.2, $+$14.6] & no & no \\
 & dual & $\mathcal{C}$ & $-$8.3 & 0.0 & $+$8.3 & [$-$4.2, $+$19.8] & no & no \\
 & ZoomClick-noPZ & $\mathcal{C}$ & $-$3.1 & $-$2.1 & $+$1.0 & [$-$8.3, $+$10.4] & no & no \\
\addlinespace
UI-Vision, sonnet$\to$opus & center & $\mathcal{T}$ & $-$5.6 & $-$2.8 & $+$2.8 & [$-$1.4, $+$6.9] & no & no \\
 & grid & $\mathcal{T}$ & $-$7.6 & $-$0.7 & $+$6.9 & [$+$0.7, $+$13.9] & no & no \\
 & context & $\mathcal{T}$ & $-$5.6 & 0.0 & $+$5.6 & [$+$0.7, $+$11.1] & no & no \\
 & dual & $\mathcal{T}$ & $-$10.4 & $-$6.2 & $+$4.2 & [$-$2.1, $+$11.1] & no & no \\
 & ZoomClick-noPZ & $\mathcal{T}$ & $-$5.6 & $-$0.7 & $+$4.9 & [$+$0.7, $+$9.7] & no & no \\
\addlinespace
ScreenSpot-Pro, haiku$\to$sonnet & center & $\mathcal{C}$ & $+$14.6 & $+$16.7 & $+$2.1 & [$-$8.3, $+$12.5] & no & no \\
 & grid & $\mathcal{C}$ & $+$20.8 & $+$14.6 & $-$6.2 & [$-$10.4, $-$2.1] & no & no \\
 & context & $\mathcal{C}$ & $+$16.7 & $+$10.4 & $-$6.2 & [$-$16.7, $+$4.2] & no & no \\
 & dual & $\mathcal{C}$ & $+$12.5 & $+$12.5 & 0.0 & [$-$22.9, $+$18.8] & no & no \\
 & ZoomClick-noPZ & $\mathcal{C}$ & $+$25.0 & $+$20.8 & $-$4.2 & [$-$14.6, $+$4.2] & no & no \\
\addlinespace
ScreenSpot-Pro, sonnet$\to$opus & center & $\mathcal{T}$ & $+$10.4 & $+$15.6 & $+$5.2 & [$-$2.1, $+$12.5] & no & no \\
 & grid & $\mathcal{T}$ & $+$11.5 & $+$14.6 & $+$3.1 & [$-$5.2, $+$11.5] & no & no \\
 & context & $\mathcal{T}$ & $+$11.5 & $+$12.5 & $+$1.0 & [$-$6.2, $+$8.3] & no & no \\
 & dual & $\mathcal{T}$ & $+$14.6 & $+$15.6 & $+$1.0 & [$-$10.4, $+$10.4] & no & no \\
 & ZoomClick-noPZ & $\mathcal{T}$ & $+$17.7 & $+$16.7 & $-$1.0 & [$-$7.3, $+$5.2] & no & no \\
\bottomrule
\end{tabular}\end{adjustbox}

\caption{Cross-model change of each skill's effect: source and target effect, $\Delta\hat\tau$ (points) with application-level 95\% interval, and TOST equivalence $p$ at $\pm1$ and $\pm5$ points; basis: pool on which both effects are measured.}
\label{tab:revdtau}
\end{table}

\begin{table}[!htbp]\centering\tiny
\begin{adjustbox}{max width=\linewidth}\begin{tabular}{ll l r rr rr}
\toprule
Domain, change & Comparison & pool & $n$ & $0{\to}1$ & $1{\to}0$ & rate $0{\to}1$ (\%) & rate $1{\to}0$ (\%) \\
\midrule
UI-Vision, haiku$\to$sonnet & direct & $\mathcal{C}$ & 96 & 16 & 4 & 16.7 & 4.2 \\
 & center & $\mathcal{C}$ & 96 & 21 & 1 & 21.9 & 1.0 \\
 & grid & $\mathcal{C}$ & 96 & 25 & 2 & 26.0 & 2.1 \\
 & context & $\mathcal{C}$ & 96 & 19 & 2 & 19.8 & 2.1 \\
 & dual & $\mathcal{C}$ & 96 & 22 & 2 & 22.9 & 2.1 \\
 & ZoomClick-noPZ & $\mathcal{C}$ & 96 & 14 & 1 & 14.6 & 1.0 \\
UI-Vision, haiku$\to$sonnet & direct & $\mathcal{T}$ & 144 & 26 & 3 & 18.1 & 2.1 \\
\addlinespace
UI-Vision, sonnet$\to$opus & direct & $\mathcal{C}$ & 96 & 7 & 4 & 7.3 & 4.2 \\
 & center & $\mathcal{C}$ & 96 & 6 & 2 & 6.2 & 2.1 \\
 & grid & $\mathcal{C}$ & 96 & 11 & 4 & 11.5 & 4.2 \\
 & context & $\mathcal{C}$ & 96 & 9 & 3 & 9.4 & 3.1 \\
 & dual & $\mathcal{C}$ & 96 & 5 & 3 & 5.2 & 3.1 \\
 & ZoomClick-noPZ & $\mathcal{C}$ & 96 & 10 & 4 & 10.4 & 4.2 \\
 & direct & $\mathcal{T}$ & 144 & 6 & 5 & 4.2 & 3.5 \\
 & center & $\mathcal{T}$ & 144 & 8 & 3 & 5.6 & 2.1 \\
 & grid & $\mathcal{T}$ & 144 & 18 & 7 & 12.5 & 4.9 \\
 & context & $\mathcal{T}$ & 144 & 10 & 1 & 6.9 & 0.7 \\
 & dual & $\mathcal{T}$ & 144 & 15 & 8 & 10.4 & 5.6 \\
 & ZoomClick-noPZ & $\mathcal{T}$ & 144 & 11 & 3 & 7.6 & 2.1 \\
\addlinespace
UI-Vision, sonnet rep.~0 vs 1 & direct & $\mathcal{T}$ & 144 & 1 & 7 & 0.7 & 4.9 \\
UI-Vision, opus rep.~0 vs 1 & direct & $\mathcal{T}$ & 144 & 1 & 2 & 0.7 & 1.4 \\
\addlinespace
ScreenSpot-Pro, haiku$\to$sonnet & direct & $\mathcal{C}$ & 48 & 6 & 2 & 12.5 & 4.2 \\
 & center & $\mathcal{C}$ & 48 & 6 & 1 & 12.5 & 2.1 \\
 & grid & $\mathcal{C}$ & 48 & 3 & 2 & 6.2 & 4.2 \\
 & context & $\mathcal{C}$ & 48 & 3 & 2 & 6.2 & 4.2 \\
 & dual & $\mathcal{C}$ & 48 & 6 & 2 & 12.5 & 4.2 \\
 & ZoomClick-noPZ & $\mathcal{C}$ & 48 & 2 & 0 & 4.2 & 0.0 \\
ScreenSpot-Pro, haiku$\to$sonnet & direct & $\mathcal{T}$ & 96 & 22 & 3 & 22.9 & 3.1 \\
\addlinespace
ScreenSpot-Pro, sonnet$\to$opus & direct & $\mathcal{C}$ & 48 & 5 & 1 & 10.4 & 2.1 \\
 & center & $\mathcal{C}$ & 48 & 1 & 0 & 2.1 & 0.0 \\
 & grid & $\mathcal{C}$ & 48 & 3 & 1 & 6.2 & 2.1 \\
 & context & $\mathcal{C}$ & 48 & 4 & 0 & 8.3 & 0.0 \\
 & dual & $\mathcal{C}$ & 48 & 6 & 5 & 12.5 & 10.4 \\
 & ZoomClick-noPZ & $\mathcal{C}$ & 48 & 3 & 1 & 6.2 & 2.1 \\
 & direct & $\mathcal{T}$ & 96 & 8 & 6 & 8.3 & 6.2 \\
 & center & $\mathcal{T}$ & 96 & 8 & 1 & 8.3 & 1.0 \\
 & grid & $\mathcal{T}$ & 96 & 9 & 4 & 9.4 & 4.2 \\
 & context & $\mathcal{T}$ & 96 & 5 & 2 & 5.2 & 2.1 \\
 & dual & $\mathcal{T}$ & 96 & 6 & 3 & 6.2 & 3.1 \\
 & ZoomClick-noPZ & $\mathcal{T}$ & 96 & 3 & 2 & 3.1 & 2.1 \\
\addlinespace
ScreenSpot-Pro, sonnet rep.~0 vs 1 & direct & $\mathcal{T}$ & 96 & 3 & 6 & 3.1 & 6.2 \\
ScreenSpot-Pro, opus rep.~0 vs 1 & direct & $\mathcal{T}$ & 96 & 5 & 5 & 5.2 & 5.2 \\
\bottomrule
\end{tabular}\end{adjustbox}

\caption{Per-task flips ($0{\to}1$, $1{\to}0$) between source and target model, for each skill's workflow and for direct clicks, and between repeated direct runs of one model, with the pool and its size.}
\label{tab:revflips}
\end{table}

\paragraph{Resolution, subset and target type.} Table~\ref{tab:revres} gives native screenshot sizes, the contract's downscale factor and target sizes per domain and role; Table~\ref{tab:revsubset} compares the retained UI-Vision roots with those removed by the per-screenshot cap; Table~\ref{tab:revtexticon} splits $\mathcal{A}$, $\mathcal{C}$ and $\mathcal{T}$ and the sealed effects by text and icon targets. ScreenSpot-Pro's maintenance pool is \pct{rev.desc.d7.sspro.composition.C.share-icon}\% icon targets against \pct{rev.desc.d7.sspro.composition.T.share-icon}\% on its sealed test, and on ScreenSpot-Pro skills help icons more than text.

\begin{table}[!htbp]\centering\scriptsize
\begin{adjustbox}{max width=\linewidth}\begin{tabular}{l l r c c c c c c}
\toprule
Domain & role & $n$ & long edge (px) & short edge (px) & MP & downscale $s$ & target short side, native (px) & in direct view (px) \\
\midrule
UI-Vision & A & 64 & 1920 (1680--2310) & 1044 (1020--1290) & 2.0 (1.8--3.2) & 0.53 (0.44--0.61) & 30.8 (24.6--44.3) & 16.3 (14.0--21.5) \\
 & C & 96 & 1920 (1773--1920) & 1044 (953--1080) & 2.0 (1.8--2.1) & 0.53 (0.53--0.58) & 26.6 (23.4--34.2) & 15.1 (12.8--17.9) \\
 & T & 144 & 1920 (1680--1920) & 1044 (1008--1080) & 2.0 (1.8--2.1) & 0.53 (0.53--0.61) & 27.6 (22.9--36.0) & 16.1 (13.1--19.2) \\
 & all & 304 & 1920 (1680--1920) & 1044 (1002--1080) & 2.0 (1.8--2.1) & 0.53 (0.53--0.61) & 27.8 (23.3--37.0) & 15.6 (13.1--19.4) \\
\addlinespace
ScreenSpot-Pro & A & 32 & 2560 (2460--3456) & 1520 (1440--2160) & 3.9 (3.5--7.5) & 0.40 (0.30--0.42) & 22.0 (17.0--29.0) & 8.0 (6.0--9.3) \\
 & C & 48 & 3840 (3840--3840) & 2160 (2160--2160) & 8.3 (7.7--8.3) & 0.27 (0.27--0.27) & 22.5 (20.0--27.2) & 6.9 (5.3--10.5) \\
 & T & 96 & 2880 (2560--3552) & 1440 (1440--1961) & 4.1 (3.7--6.1) & 0.36 (0.29--0.40) & 24.0 (19.8--31.2) & 8.8 (6.8--10.4) \\
 & all & 176 & 3456 (2560--3840) & 1664 (1440--2160) & 4.7 (3.7--7.9) & 0.30 (0.27--0.40) & 23.0 (19.0--30.0) & 8.1 (6.6--10.4) \\
\bottomrule
\end{tabular}\end{adjustbox}

\caption{Native screenshot size, megapixels, downscale factor $s$ of the direct view under the image contract, and target shorter side natively and in the direct view (median [IQR]).}
\label{tab:revres}
\end{table}

\begin{table}[!htbp]\centering\scriptsize
\begin{adjustbox}{max width=\linewidth}\begin{tabular}{l r r r c c c}
\toprule
UI-Vision group & roots & screenshots & apps & icon (\%) & target short side, native px & share from screenshots with $>2$ roots (\%) \\
\midrule
all valid annotated roots & 1744 & 1164 & 83 & 78.8 & 27.2 (22.3--34.3) & 25.3 \\
removed by the 2-per-screenshot cap & 183 & 129 & 45 & 86.9 & 26.4 (22.4--32.8) & 100.0 \\
kept (census) & 1561 & 1164 & 83 & 77.8 & 27.3 (22.3--34.6) & 16.5 \\
$\mathcal{A}\cup\mathcal{C}\cup\mathcal{T}$ before reduction & 943 & 701 & 44 & 79.4 & 27.2 (22.2--33.6) & 15.7 \\
used ($\mathcal{A}\cup\mathcal{C}\cup\mathcal{T}$) & 304 & 281 & 44 & 77.3 & 27.8 (23.3--37.0) & 15.1 \\
\bottomrule
\end{tabular}\end{adjustbox}

\caption{UI-Vision element-grounding roots before and after the per-screenshot cap.}
\label{tab:revsubset}
\end{table}

\begin{table}[!htbp]\centering\scriptsize
\begin{adjustbox}{max width=\linewidth}\begin{tabular}{l ccc ccc}
\toprule
 & \multicolumn{3}{c}{agree/stable share (\%)} & \multicolumn{3}{c}{agree/stable $F-R$, catalogue pooled (pt)} \\ \cmidrule(lr){2-4}\cmidrule(lr){5-7}
Setting & frozen 2\% diag. & $\le$ target diag. & both in box & frozen & $\le$ target diag. [95\% CI] & both in box \\
\midrule
UI-Vision, haiku$\to$sonnet, $\mathcal{C}$ & 70.6 & 71.9 & 58.1 & $+$1.8 & $+$2.5 [0.0, $+$5.4] & $-$0.4 \\
UI-Vision, sonnet$\to$opus, $\mathcal{C}$ & 87.0 & 87.0 & 71.6 & $+$0.6 & $+$0.9 [$-$2.8, $+$4.7] & $-$2.5 \\
UI-Vision, sonnet$\to$opus, $\mathcal{T}$ & 88.9 & 89.6 & 76.6 & $-$0.8 & $-$1.4 [$-$4.0, $+$0.8] & $-$2.3 \\
ScreenSpot-Pro, haiku$\to$sonnet, $\mathcal{C}$ & 76.0 & 74.5 & 56.8 & $+$1.4 & $+$2.8 [$-$0.6, $+$6.1] & $-$0.9 \\
ScreenSpot-Pro, sonnet$\to$opus, $\mathcal{C}$ & 83.3 & 78.1 & 68.8 & $-$1.2 & $-$2.0 [$-$4.9, 0.0] & $-$3.8 \\
ScreenSpot-Pro, sonnet$\to$opus, $\mathcal{T}$ & 87.0 & 85.9 & 71.9 & $+$0.9 & $+$0.9 [$-$1.2, $+$2.9] & $-$0.7 \\
\bottomrule
\end{tabular}\end{adjustbox}

\caption{Agree/stable share and pooled residual under the frozen 2\%-of-diagonal cut-off and under target-size-relative cut-offs.}
\label{tab:revrestrat}
\end{table}

\begin{table}[!htbp]\centering\tiny
\begin{adjustbox}{max width=\linewidth}\begin{tabular}{ll l rr c rr c}
\toprule
 & & & \multicolumn{3}{c}{text} & \multicolumn{3}{c}{icon} \\ \cmidrule(lr){4-6}\cmidrule(lr){7-9}
Domain & target & Skill & $n$ & apps & $\hat\tau$ (pt) [95\% CI] & $n$ & apps & $\hat\tau$ (pt) [95\% CI] \\
\midrule
UI-Vision & sonnet & center & 30 & 15 & $-$9.4 [$-$23.9, 0.0] & 114 & 23 & $-$3.6 [$-$9.9, $+$1.7] \\
 &  & grid & 30 & 15 & $-$8.3 [$-$23.3, 0.0] & 114 & 23 & $-$6.4 [$-$13.2, $+$0.7] \\
 &  & context & 30 & 15 & $-$9.4 [$-$23.9, 0.0] & 114 & 23 & $-$3.9 [$-$9.4, $+$0.4] \\
 &  & dual & 30 & 15 & $-$6.7 [$-$20.0, 0.0] & 114 & 23 & $-$11.4 [$-$17.1, $-$5.9] \\
 &  & ZoomClick-noPZ & 30 & 15 & $-$6.7 [$-$20.0, 0.0] & 114 & 23 & $-$5.9 [$-$12.0, $-$1.0] \\
\addlinespace
 & opus & center & 30 & 15 & 0.0 [0.0, 0.0] & 114 & 23 & $-$2.8 [$-$8.0, $+$1.2] \\
 &  & grid & 30 & 15 & $+$3.3 [0.0, $+$10.0] & 114 & 23 & $-$1.6 [$-$5.5, $+$2.2] \\
 &  & context & 30 & 15 & 0.0 [0.0, 0.0] & 114 & 23 & $+$0.1 [$-$2.9, $+$3.2] \\
 &  & dual & 30 & 15 & $+$6.7 [0.0, $+$20.0] & 114 & 23 & $-$7.7 [$-$15.2, $-$1.2] \\
 &  & ZoomClick-noPZ & 30 & 15 & 0.0 [$-$20.0, $+$20.0] & 114 & 23 & $-$0.6 [$-$4.3, $+$3.0] \\
\addlinespace
ScreenSpot-Pro & sonnet & center & 68 & 12 & $+$9.3 [$+$1.4, $+$20.5] & 28 & 10 & $+$16.7 [$-$5.0, $+$41.0] \\
 &  & grid & 68 & 12 & $+$5.1 [$-$7.9, $+$19.4] & 28 & 10 & $+$25.0 [$+$2.0, $+$52.0] \\
 &  & context & 68 & 12 & $+$8.6 [$-$2.7, $+$22.6] & 28 & 10 & $+$20.0 [$-$1.7, $+$45.0] \\
 &  & dual & 68 & 12 & $+$10.7 [$-$2.8, $+$27.0] & 28 & 10 & $+$23.3 [$+$0.7, $+$48.7] \\
 &  & ZoomClick-noPZ & 68 & 12 & $+$13.8 [$+$4.0, $+$25.9] & 28 & 10 & $+$25.3 [$+$8.3, $+$46.7] \\
\addlinespace
 & opus & center & 68 & 12 & $+$10.2 [$+$2.8, $+$19.4] & 28 & 10 & $+$27.0 [$+$8.7, $+$48.7] \\
 &  & grid & 68 & 12 & $+$11.4 [$+$3.5, $+$20.8] & 28 & 10 & $+$21.7 [$+$4.0, $+$43.3] \\
 &  & context & 68 & 12 & $+$9.2 [$+$2.4, $+$18.1] & 28 & 10 & $+$18.7 [$-$4.7, $+$43.0] \\
 &  & dual & 68 & 12 & $+$10.0 [$+$2.4, $+$19.0] & 28 & 10 & $+$25.7 [$+$6.7, $+$48.0] \\
 &  & ZoomClick-noPZ & 68 & 12 & $+$10.0 [$+$2.4, $+$19.0] & 28 & 10 & $+$29.0 [$+$10.7, $+$50.0] \\
\midrule
\multicolumn{9}{l}{UI-Vision icon share (\%): $\mathcal{A}$ 64.1 (41/64), $\mathcal{C}$ 83.3 (80/96), $\mathcal{T}$ 79.2 (114/144)} \\
\multicolumn{9}{l}{ScreenSpot-Pro icon share (\%): $\mathcal{A}$ 50.0 (16/32), $\mathcal{C}$ 54.2 (26/48), $\mathcal{T}$ 29.2 (28/96)} \\
\bottomrule
\end{tabular}\end{adjustbox}

\caption{Text vs.\ icon targets: composition of each role and sealed effect $\hat\tau$ by type (points; small cells).}
\label{tab:revtexticon}
\end{table}

\begin{table}[!htbp]\centering\scriptsize
\begin{adjustbox}{max width=\linewidth}\begin{tabular}{l rrr rrr}
\toprule
 & \multicolumn{3}{c}{UI-Vision} & \multicolumn{3}{c}{ScreenSpot-Pro} \\ \cmidrule(lr){2-4}\cmidrule(lr){5-7}
Job & req./task & MP/task & $\times$ direct & req./task & MP/task & $\times$ direct \\
\midrule
direct & 1.00 & 0.59 & 1.00 & 1.00 & 0.58 & 1.00 \\
replay (reader only) & 1.00 & 0.43 & 0.74 & 1.00 & 0.53 & 0.91 \\
center workflow & 2.00 & 1.12 & 1.92 & 2.00 & 1.18 & 2.03 \\
grid workflow & 3.69 & 2.99 & 5.10 & 3.85 & 3.73 & 6.38 \\
context workflow & 2.00 & 1.35 & 2.31 & 2.00 & 1.41 & 2.42 \\
dual workflow & 3.01 & 0.96 & 1.64 & 3.01 & 1.04 & 1.78 \\
ZoomClick-noPZ workflow & 2.99 & 1.36 & 2.32 & 2.98 & 1.66 & 2.84 \\
\bottomrule
\end{tabular}\end{adjustbox}

\caption{Image pixels sent per task relative to the direct request (and absolute megapixels for direct), per skill and operation, on the maintenance pool.}
\label{tab:revpixels}
\end{table}

\paragraph{Audit checks that a re-run needs.} Protocol-essential: a fresh session per request, a check that the returned model identifier equals the requested one, byte hashes of every view the model received, a read-scope audit of every tool call, and frozen decisions hashed before any sealed request. Harness-specific: the JPEG re-encoding guard, the zero-read rule and the pre-launch check that every file is local, which our harness needed because of its interface and of storage eviction. In our run \cnt{all.records.status.harness-failed} infrastructure failures were re-queued against \cnt{all.records.status.ok} completed requests.

\end{document}